\documentclass[10pt,journal,compsoc]{IEEEtran}
\ifCLASSOPTIONcompsoc
  \usepackage[nocompress]{cite}
\else
  \usepackage{cite}
\fi
\ifCLASSINFOpdf
\else
\fi

\usepackage{microtype}                 
\PassOptionsToPackage{warn}{textcomp}  
\usepackage{textcomp}                  
\usepackage{mathptmx}                  
\usepackage{times}                     
\usepackage{cite}                      
\usepackage{tabu}                      
\usepackage{booktabs}                  
\usepackage{amsmath}
\usepackage{amsfonts}
\usepackage{dsfont}
\usepackage{relsize}
\usepackage{mathtools}
\usepackage[ruled,linesnumbered]{algorithm2e}
\usepackage{makecell}
\usepackage{multirow}
\usepackage{enumitem}

\usepackage{microtype}                 
\PassOptionsToPackage{warn}{textcomp}  
\usepackage{textcomp}                  
\usepackage{mathptmx}                  
\usepackage{amssymb}
\usepackage{times}                     
\usepackage{cite}                      
\usepackage{tabu}                      
\usepackage{booktabs}   
\usepackage{comment}
\usepackage{soul}
\usepackage{float}
\usepackage{multirow}
\usepackage{array}
\newcolumntype{L}{>{\centering\arraybackslash}m{3cm}}

\usepackage{fancyhdr} 
\usepackage[ruled,linesnumbered]{algorithm2e}
\usepackage{multicol}
\usepackage[T1]{fontenc}
\usepackage{makecell}
\usepackage[normalem]{ulem}
\usepackage{xcolor}

\usepackage[colorlinks,urlcolor=blue,linkcolor=blue,citecolor=blue]{hyperref}
\usepackage{enumitem}
\usepackage{microtype}                 
\PassOptionsToPackage{warn}{textcomp}  
\usepackage{textcomp}                  
\usepackage{mathptmx}                  
\usepackage{amssymb}
\usepackage{times}                     
\usepackage{cite}                      
\usepackage{tabu}                      
\usepackage{booktabs}   
\usepackage{comment}
\usepackage{soul}
\usepackage{float}
\usepackage{multirow}
\usepackage{array}
\usepackage{subcaption}
\usepackage{wrapfig}
\usepackage[export]{adjustbox}
\usepackage{hyperref}
\hypersetup{
    colorlinks=true,
}

\usepackage{fancyhdr} 
\usepackage[ruled,linesnumbered]{algorithm2e}
\usepackage{multicol}
\usepackage[T1]{fontenc}
\usepackage{makecell}
\usepackage[normalem]{ulem}
\usepackage{xcolor}

\usepackage[medium,compact]{titlesec} 
\begin{document}

\newcommand{\ganName}{the self-challenging mechanism}

\newcommand{\modelName}{SmartGD}
\newcommand{\dis}{\Phi_\mathrm{dis}}
\newcommand{\gen}{\Phi_\mathrm{gen}}
\newcommand{\Xr}[1]{\mathbf{X}_r^{(#1)}}
\newcommand{\Xf}[1]{\mathbf{X}_f^{(#1)}}
\newcommand{\XrG}[2]{\mathbf{X}_{r,G_{#2}}^{(#1)}}
\newcommand{\XfG}[2]{\mathbf{X}_{f,G_{#2}}^{(#1)}}
\newcommand{\red}{\textcolor{black}}
\newcommand{\revise}{\textcolor{black}}
\newcommand{\minorrevise}{\textcolor{black}}

%



\title{\modelName: A GAN-Based Graph Drawing Framework for Diverse Aesthetic Goals}
%
%
%
%

\author{Xiaoqi~Wang,
        Kevin~Yen,
        Yifan~Hu,
        and~Han-Wei~Shen
\IEEEcompsocitemizethanks{\IEEEcompsocthanksitem Xiaoqi Wang and Han-Wei Shen are with The Ohio State University\protect\\
E-mail: wang.5502@osu.edu, shen.94@osu.edu
\IEEEcompsocthanksitem Kevin Yen and Yifan Hu are with Yahoo! Research\protect\\
Email: kevinyen@yahooinc.com, yifanh@gmail.com}
}

%
%

\markboth{Journal of \LaTeX\ Class Files,~Vol.~14, No.~8, August~2022}%
{Shell \MakeLowercase{\textit{et al.}}: Bare Demo of IEEEtran.cls for Computer Society Journals}
%



\IEEEtitleabstractindextext{%
\begin{abstract}
\minorrevise{While a multitude of studies have been conducted on graph drawing, many existing methods only focus on optimizing a single aesthetic aspect of graph layouts, which can lead to sub-optimal results.} There are a few existing methods \minorrevise{that have attempted} to develop a flexible solution for optimizing different aesthetic aspects measured by different aesthetic criteria. Furthermore, thanks to the significant advance in deep learning techniques, several deep learning-based layout methods were proposed recently. These methods have demonstrated the advantages of \minorrevise{deep learning approaches} for graph drawing. However, none of these existing methods can be directly applied to optimizing non-differentiable criteria without special accommodation. \red{In this work, we propose a novel Generative Adversarial Network (GAN) based deep learning framework for graph drawing, called \modelName, which can optimize different quantitative aesthetic goals, \minorrevise{regardless of} their differentiability. To demonstrate the effectiveness and efficiency of \modelName, we \minorrevise{conducted} experiments on minimizing stress, minimizing edge crossing, maximizing crossing angle, maximizing shape-based metrics, and a combination of multiple aesthetics.} Compared with several popular graph drawing algorithms, the experimental results show that \modelName\ achieves good performance both quantitatively and qualitatively.\end{abstract}

}

\maketitle

\IEEEdisplaynontitleabstractindextext

%
\IEEEpeerreviewmaketitle

\IEEEraisesectionheading{\section{Introduction}\label{sec:introduction}}

%
%
%
%
\IEEEPARstart{A}graph is a mathematical structure that can be used to model networks (\minorrevise{e.g.,} social networks, transportation networks) in many different applications. \red{There are several common aesthetic criteria formulated by researchers to evaluate the readability of graph layouts\cite{purchase1}. They can be classified into two categories: differentiable criteria (\minorrevise{e.g.,} stress, the uniformity of edge length, angles formed by pairs of incident edges, etc.) and non-differentiable criteria (\minorrevise{e.g.,} the \minorrevise{number of edge crossings}, angles formed by pairs of crossing edges, shape-based metrics\cite{shape-metric}, etc.). Mathematically speaking, any criterion that is differentiable everywhere \minorrevise{with respect to} the node positions is \minorrevise{considered a} differentiable criterion. It is worth pointing out that each of these criteria, regardless of their differentiability, is an important factor in evaluating graph layouts because \minorrevise{each criterion focuses on measuring distinct aesthetic aspects}. There is no consensus on which criterion is the most important\cite{ Tim-user-study}.}

\red{In terms of generating straight-line drawings for general graphs, most existing works focus on optimizing a single aesthetic criterion. \revise{Developing layout methods that can flexibly optimize any aesthetic criteria is still very challenging.} A few methods (i.e., DeepGD\cite{deepgd}, SPX\cite{spx}, and $\mathrm{GD}^2$\cite{sgd2}) have been proposed recently with the purpose of optimizing different aesthetics, or even a combination of aesthetics. Nonetheless, they all have their own limitations: SPX can only optimize a limited set of criteria; DeepGD and $\mathrm{GD}^2$ are more flexible than SPX, but they are designed to optimize differentiable aesthetic criteria. For non-differentiable aesthetic criteria, DeepGD and $\mathrm{GD}^2$ require manually designed differentiable surrogate functions to approximate \minorrevise{these criteria.}} Developing a unified layout method for straight-line graph drawing with diverse aesthetic goals, \minorrevise{including non-differentiable ones,} remains a non-trivial task.   



Recently, thanks to the significant advances in deep learning techniques, several methods have shown success in generating straight-line graph drawing\cite{deep-drawing,deepgd,kwon-ma-2020,dnn}. From the initial studies that use deep learning-based methods for producing graph layouts, it becomes evident that the deep learning approach, by its nature, has several advantages \minorrevise{over traditional approaches.} First of all, while training a deep learning model can be time-consuming, once trained, applying the trained model to a new graph can be computationally more efficient than the traditional layout methods. Specifically, the graph layouts can be generated with just a single forward pass over the model, while many traditional methods\cite{neato,Bekos-xangle,sgd2,kamada_kawai_1989} are iterative algorithms that may require a large number of iterations. Secondly, \minorrevise{the deep learning approaches} are data-driven, which makes it possible for them to achieve a variety of goals as long as proper training data are available, even for tasks in graph drawing that have not been addressed before by traditional approaches. For example, Kwon et al.~\cite{kwon-ma-2020} \minorrevise{proposed} a generative model to learn a latent space for smooth transitions between existing layouts.
\red{Lastly, the deep learning approaches learn to draw graphs \minorrevise{based on general knowledge} extracted from a large collection of graphs, while \minorrevise{most traditional graph drawing approaches} consider only \minorrevise{a specific instance of a graph.} Hence, the deep learning approaches employ more general strategies to circumvent \minorrevise{local minima when optimizing} graph layouts.} 
Because of these advantages, deep learning for graph drawing is a direction that has the potential to bring about major quality and performance improvements, similar to what has been seen in other applications such as Natural Language Processing and Computer Vision. In the existing literature, there is only one deep learning layout method focusing on optimizing graph layouts toward different aesthetic goals, called DeepGD, which generates layouts by minimizing a loss function that is composed of multiple aesthetic criteria. \red{However, the limitation of DeepGD is that the aesthetic criteria to be optimized must be differentiable. To date, \minorrevise{the development of deep learning approaches that can work for both differentiable and non-differentiable criteria is still an under-explored area.}} 

\red{In this paper, we perform further studies along the direction of drawing graphs via \minorrevise{deep learning methods.}
We propose a novel Generative Adversarial Network (GAN)\cite{sgan} framework, called \modelName, to generate straight-line graph drawings. In essence, the proposed framework is a general solution for optimizing diverse quantitative aesthetic goals, irrespective of their differentiability. If there are multiple criteria of interest, \modelName\ can also flexibly optimize the layouts toward \minorrevise{a combination of them} based on pre-defined importance for each criterion. To optimize graph layouts toward an optimization goal without the need to \minorrevise{mathematically define aesthetic metrics}, we allow SmartGD to learn from high-quality layouts generated by itself and keep challenging the self-generated layouts during training via the self-challenging mechanism. Compared to existing works focusing on optimizing different criteria\cite{spx,sgd2,deepgd}, this work addresses the unique challenge of optimizing non-differentiable criteria without the need to manually define a differentiable surrogate function~\cite{sgd2} for them.}

To validate the proposed approach, we conduct experiments on generating graph layouts with regard to different criteria, including minimizing edge crossing, maximizing crossing angle, minimizing stress, \red{maximizing shape-based metrics}, and optimizing a combination of multiple aesthetics. The effectiveness and efficiency of \modelName\ are evaluated quantitatively and qualitatively against several widely used graph drawing algorithms as benchmarks. \revise{The experimental results show that our method achieves good performance in terms of all four aesthetics and a combination of seven aesthetics compared with the benchmark algorithms, both quantitatively and qualitatively. }

    


\section{Related Work}

\subsection{Graph Drawing}
Since 1963, a multitude of graph drawing algorithms has been proposed. In order to evaluate the goodness of graph layouts, several commonly agreed aesthetic criteria (e.g., number of edge crossing, minimum crossing angle, and node occlusion)\cite{purchase1} are formulated by researchers. Extensive user studies have shown that these criteria are highly correlated with human preference regarding graph layouts\cite{Tim-user-study}. To be specific, each aesthetic criterion emphasizes a single aspect of aesthetics, and some criteria contradict each other\cite{haleem-huamin}. Until now, there is no general agreement about which criterion is the most effective one to measure human preference. 

In general, one of the most popular ways to generate straight-line drawings for general graphs is through energy-based algorithms\cite{kobourov_2013}, for example, by minimizing stress\cite{neato,kamada_kawai_1989,zheng-gd2}. In addition to stress, there are several layout methods designed to optimize edge crossings or crossing angles in the straight-line drawing. However, compared with stress minimization, they are still less explored mainly because optimizing these two aesthetics is an NP-hard problem\cite{Bekos-xangle,xing-heuristic}. Specifically, to maximize the crossing angle, a force-directed-based algorithm \cite{Argyriou} and a heuristic-based algorithm \cite{Bekos-xangle} \minorrevise{was proposed}. To minimize the edge crossing in straight-line drawings, a heuristic-based algorithm \cite{xing-heuristic} is proposed and has shown that it can produce fewer edge crossings than the energy-based algorithms. However, they all focus on optimizing a single aesthetic criterion. Unlike these methods, \modelName\ is more flexible because it can optimize any quantitative criteria.

It is widely recognized that a good graph layout often complies with multiple aesthetics simultaneously\cite{didimo}. There are several researches\cite{sgd2,spx} conducted to optimize different aesthetic criteria or a combination of criteria. For example, Stress-Plus-X\cite{spx} (SPX) attempts to optimize stress together with an "X" criterion by combining the stress minimization with the penalty terms representing the "X" criteria (i.e., edge crossings, crossing angle, and upwardness). However, it has a limitation such that it can only be applied to a limited set of criteria. Then, Ahmed et al~\cite{sgd2}, propose a more flexible framework called $\mathrm{GD}^2$, which utilizes stochastic gradient descent to optimize the graph layout with respect to any differentiable criteria, including a combination of multiple differentiable criteria. However, it cannot be directly generalized to non-differentiable aesthetic criteria without special accommodation. Therefore, they carefully design hand-crafted differentiable surrogate functions to approximate some non-differentiable criteria such that stochastic gradient descent can be applied.
It is worth mentioning that \modelName\ is also capable of optimizing a combination of multiple criteria, but is more general than these two methods because \modelName\ can optimize any non-differentiable criterion without \minorrevise{the need for manually defined differentiable surrogate functions}.

\subsection{Deep Learning Approaches for Graph Drawing}
In recent years, deep learning techniques have achieved state-of-the-art performance in different applications, such as Natural Language Processing and Computer Vision. Thanks to the rapid advancement of deep learning, researchers have successfully developed multiple deep learning-based graph drawing methods\cite{dnn,gnn-gd,kwon-ma-2020,deep-drawing,deepgd}. Specifically, Wang et al.~\cite{deep-drawing} propose an LSTM-based graph drawing model called DeepDrawing to visualize graphs in a similar layout fashion as the training data. However, since DeepDrawing encodes the graph structure information using an adjacency vector with a fixed length $k$ for each node, only the connectivity information between the current node and $k$ other nodes is accessible to DeepDrawing. As a result, DeepDrawing is unable to capture the global graph topology information, which makes it difficult to draw unseen graphs that have different topological characteristics than the graphs in the training data. Besides, $(DNN)^2$\cite{dnn} employs Graph Convolution Network to generate layouts by optimizing stress and tsNET\cite{tsnet}. DeepGD\cite{deepgd} is proposed to generate the optimal layout according to an aesthetic criterion or a combination of criteria. However, similar to $\mathrm{GD}^2$, DeepGD is only applicable to optimizing differentiable criteria. Later on, Tiezzi et al.\cite{gnn-gd} propose to use a neural network as a differentiable surrogate for computing non-differentiable criterion (i.e., edge crossing) such that this surrogate network can serve as a guidance to supervise a drawing network to draw graphs. Their limitation is that the quality of layouts generated by the drawing network largely depends on how well the surrogate network can estimate the number of edge crossings.

\red{Compared to the existing deep learning graph drawing models, \modelName\ is the first GAN-based framework for general graph drawing. This framework is a versatile solution for optimizing the quantitative aesthetic goals (even if they are non-differentiable), without the need of training a surrogate model to approximate the non-differentiable criteria or specifically designing a surrogate function for each individual non-differentiable criterion. Additionally, unlike DeepDrawing, both the local neighborhood information and the global graph structure are captured by \modelName\ during the learning progress. }  

\subsection{Generative Adversarial Networks}
\label{sec:gan-literature}
Generative Adversarial Networks (GAN) are designed to learn a generative distribution that can ultimately approximate the distribution of real data\cite{rgan,conditional-gan}. In 2014, the first GAN~\cite{sgan} emerged to generate fake images which looked like real images. Inspired by the great success of mimicking examples provided in the training data, GANs are adopted to tackle other problems such as super-resolution. Later on, a conditional version of GAN \cite{conditional-gan} is proposed to learn a conditional generative distribution, by conditioning on some additional information. \red{Besides, to alleviate the overfitting issue of the discriminator, DeceiveD\cite{DeceiveD} proposes an adaptive pseudo augmentation method to adaptively present some fake data as real data with a probability dynamically adjusted based on the degree of overfitting. In other words, if the discriminator has a tendency to overfit, the fake data are more likely to be presented as real data to suppress the confidence of the discriminator in distinguishing reals and fakes. } It is worth mentioning that there are many previous works on the design of adversarial loss. For example, WGAN\cite{wgan} is proposed to use the Wasserstein distance to estimate the distance between the generative distributions and distribution of real data to encourage faster convergence. RGAN\cite{rgan} is designed to estimate the relativistic difference between two distributions and thus can generate fake data with better quality than WGAN.

In this work, by taking advantage of the conditional RGAN, \modelName\ learns a generative layout distribution conditioned on the graph structure. \red{In order to generate optimal layouts regarding the chosen aesthetic criteria, \modelName\ will learn from the layouts generated by itself that have better aesthetic scores and continue to challenge the self-generated layouts via \ganName. Specifically, \ganName\ resembles the adaptive pseudo augmentation in DeceiveD\cite{DeceiveD} because it also presents fake data as real data in some cases. However, \ganName\ will only present fakes as reals when the fakes perform better with respect to the aesthetic criteria than the reals, while DeceiveD presents fakes as reals with a probability adjusted based on the degree of overfitting. Besides, a work\cite{sc-gan-chemical} published in the Journal of Cheminformatics also presents a similar idea as \ganName. The difference is that \cite{sc-gan-chemical} attempts to explore novel molecules which are more likely to be a new drug, while our work aims at designing graph layouts with diverse aesthetic goals.} 

\section{\modelName}
\label{sec:method}
In this paper, we propose a general graph drawing framework that can optimize the layouts of graphs toward diverse aesthetic goals. \red{To be specific, given a quantitative aesthetic goal, \modelName\ can optimize the generated layouts toward this quantitative goal regardless of whether the criteria are differentiable or not. We accomplish our goal by developing a novel GAN-based deep learning model, which is illustrated in detail in this section.}

Let $G$ = ($\mathbf{V}$, $\mathbf{E}$) be a graph, where $\mathbf{V}$ is a set of $N$ nodes, \minorrevise{ and $\mathbf{E}$ is a set of $M$ edges.} A graph $G$ can be represented by an adjacency matrix $\mathbf{A}$, where $a_{ij} = 1$ indicates there exists an edge between nodes $i$ and $j$, while $a_{ij} = 0$ suggests the opposite. The graph layout is denoted as $\mathbf{X} \in \mathbb{R}^{N \times 2}$, where the \minorrevise{$i^{\mathrm{th}}$ row} $\mathbf{X}_i$ is a 2-dimensional position vector for node $i$. In the following sections, the good layout example \minorrevise{and the generated layout} are represented by $\mathbf{X}_r$ and $\mathbf{X}_f$, respectively ("r" for {\bf r}eal and "f" for {\bf f}ake). \red{We denote the quantitative \minorrevise{criterion function} to be optimized as $\lambda(\mathbf{X},G)$.}


\subsection{Conditional RGAN with Self-Challenging Mechanism}
\label{sec:method}
\red{The basis of \modelName\ is a conditional\cite{conditional-gan} RGAN\cite{rgan} model which allows us to generate layouts based on a given set of good layout examples. In order to optimize graph layouts further toward the desired aesthetic goals, \ganName\ is added on top of the conditional RGAN. In this section, we explain the conditional RGAN and \ganName\ respectively in detail.}

\begin{figure}[htbp!]
    \setlength{\abovecaptionskip}{-1pt}
    \begin{center}
        \includegraphics[width=1\linewidth]{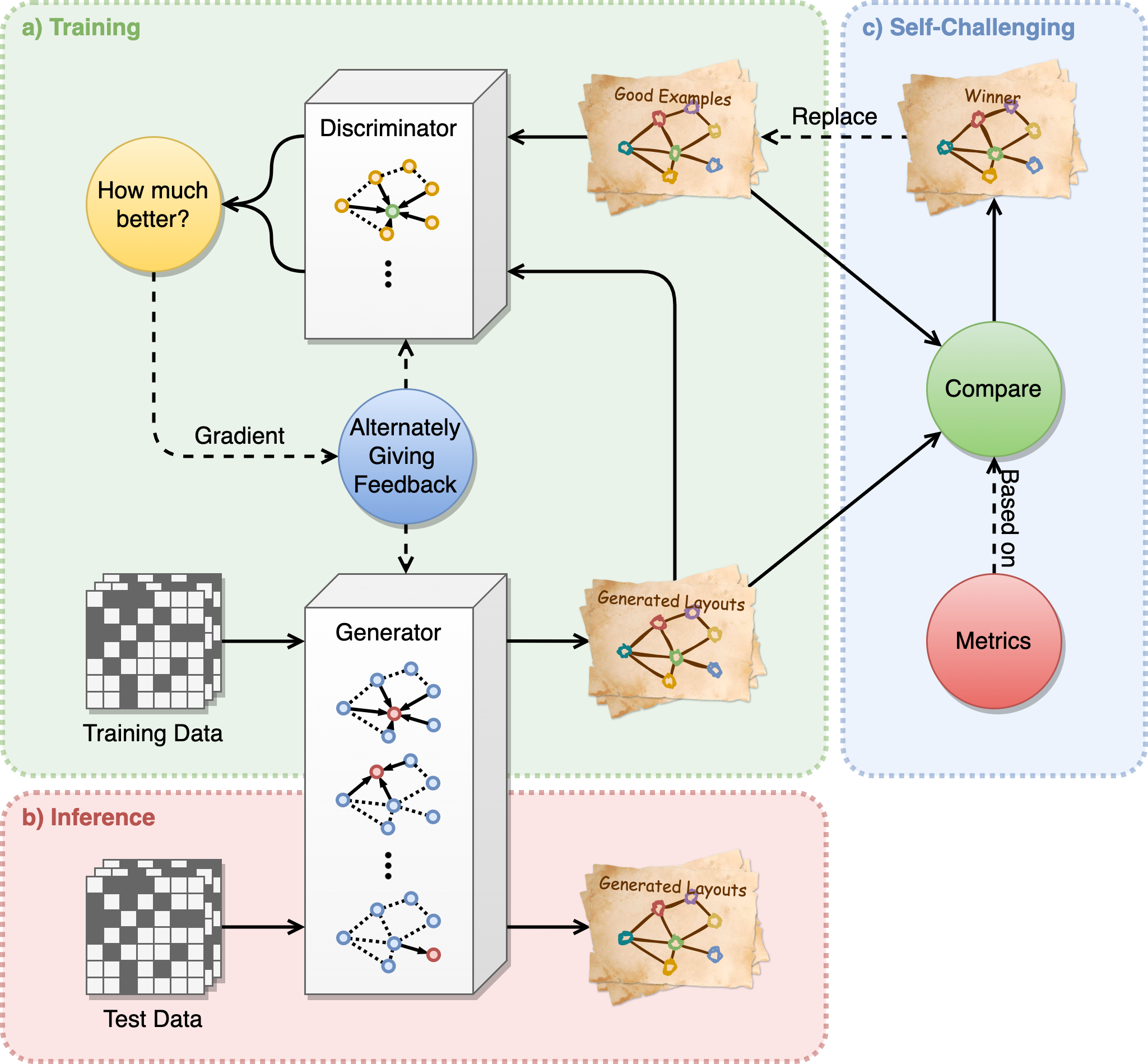}
    \end{center}
    \caption{The high-level overview of \modelName. Component (a) sketches the training procedure of the GAN-based model. The self-challenging mechanism is explained in component (c), which is applied only when the optimization criterion is given. Component (b) describes the inference procedure for drawing unseen graphs.}
    \label{fig:framework}
\end{figure}

\red{{\bf Conditional RGAN.} \minorrevise{The main goal of employing conditional RGAN is to learn how to draw layouts} by imitating a given set of good layout examples. In the model architecture, there are two sub-models: the generator network $\Phi_\mathrm{gen}$ and the discriminator network $\Phi_\mathrm{dis}$ (see \autoref{fig:framework}a). The generator is responsible for generating layouts that are as similar as possible to a given set of good layout examples, and the discriminator estimates the goodness of \minorrevise{the generated layouts}. Mathematically, the generator attempts to learn the generative distribution $\mathbb{Q}(\mathbf{X}|G)$ to approximate the distribution of good layout examples $\mathbb{P}(\mathbf{X}|G)$. Namely, the generator tries to imitate the good layouts $\mathbf{X}_r$ and aims at making the discriminator believe that the fake layouts $\mathbf{X}_f$ generated by the generator are better than the good layouts $\mathbf{X}_r$. The discriminator is responsible for correctly distinguishing $\mathbf{X}_f$ \minorrevise{from} $\mathbf{X}_r$ such that the generator will have the motivation to improve further. Therefore, \minorrevise{their responsibilities are adversarial} to some extent, but they share a common goal of helping the generator learn better. }

\red{{\bf Self-Challenging Mechanism.} 
\minorrevise{A standard conditional RGAN proves insufficient to achieve our ultimate goal of learning: draw optimal layouts based on certain aesthetic goals.} The primary reason for this is the \minorrevise{difficulty of} collecting optimal layouts with respect to criterion $\lambda$, \minorrevise{which serve as} good layout examples $\mathbf{X}_r$ for training \modelName.
Hence, if the collected good layout examples are \minorrevise{sub-optimal}, solely imitating them, as what the vanilla conditional RGAN does, will constrain the quality of the generated layouts.
Therefore, to go beyond the quality of the collected examples, we allow \modelName\ to learn from better-quality layouts generated by itself and keep challenging the highest standard in an adversarial setting. Specifically, for each layout the generator generates, we evaluate it by computing $\lambda(\mathbf{X}_f,G)$ and compare it with the current good layout example $\lambda(\mathbf{X}_r^*,G)$ (see \autoref{fig:framework}b). If the generated layout $\mathbf{X}_f$ is better than the current good layout example $\mathbf{X}_r^*$ given the criterion $\lambda$, the good layout collection will be updated by replacing that example with the newly generated layout $\mathbf{X}_f$. In this way, the generator is challenging itself by learning from the best available layouts so far and continuing to improve. Therefore, the distribution of good layout examples $\mathbb{P^*}(X|G)$ is dynamically changing and $\lambda(\mathbf{X}_r^*,G)$ is continuously improved during the training stage. More importantly, $\lambda$ can be any quantitative criterion even if it is not differentiable because $\lambda$ just serves as a guidance for replacing the good layout examples $\mathbf{X}_r^*$ \minorrevise{without the need for gradient back-propagation.} Lastly, since the essence of this mechanism is challenging the self-generated layouts in an adversarial setting, we call it \ganName. We note that although the idea is similar to the data augmentation techniques proposed in \cite{DeceiveD} and \cite{sc-gan-chemical} (see Section~\ref{sec:gan-literature}), to the best of our knowledge, this is the first time that this idea is adopted for graph drawing and shown highly effective. }

{\bf Loss Function.} The generator will take a graph $G$ as input and generate the corresponding layout $\mathbf{X}_f$. The discriminator predicts a goodness score $\dis\left(\mathbf{X}|G\right)$ for any input layout $\mathbf{X}$. The adversarial loss is defined as the following:

\begin{equation}
\begin{small}
\begin{aligned}
\label{equ:our-gan}
L_{\dis} &= -\mathbb{E}_{\mathbf{X}_r^* \sim \mathbb{P}^*(\mathbf{X}|G),\mathbf{X}_f \sim  \mathbb{Q}_\lambda(\mathbf{X}|G)} \left[\log \sigma \left( \dis\left(\mathbf{X}_r^*|G\right) - \dis\left(\mathbf{X}_f|G\right) \right) \right] \\
L_{\gen} &= -\mathbb{E}_{\mathbf{X}_r^* \sim \mathbb{P}^*(\mathbf{X}|G),\mathbf{X}_f \sim  \mathbb{Q}_\lambda(\mathbf{X}|G)} \left[\log \sigma \left( \dis\left(\mathbf{X}_f|G\right) - \dis\left(\mathbf{X}_r^*|G\right) \right) \right]
\end{aligned}
\end{small}
\end{equation}
where $\sigma$ is the sigmoid function,  $\mathbb{P}^*(\mathbf{X}|G)$ is \minorrevise{the dynamic good layout distribution, and $\mathbf{X}_r^*$ denotes the current good layout examples.} In the discriminator loss, $(\dis\left(\mathbf{X}_r|G\right) - \dis\left(\mathbf{X}_f|G\right))$ indicates how much the good layout example $\mathbf{X}_r$ is better than the generated layout $\mathbf{X}_f$ in \minorrevise{the eyes of the discriminator}. Thus, the discriminator will be trained to maximize the log probability that the good layout example is better than the generated layout in the belief of the discriminator. Similarly, the generator will be trained to maximize the log probability that the generated layout $\mathbf{X}_f$ is better than the good layout example $\mathbf{X}_r$ in \minorrevise{the belief of the discriminator}. 

\red{At the end of the training, the generator generates a layout distribution that approximates the distribution of final good layout examples $\mathbf{X}_r^*$ well. Given that $\mathbf{X}_r^*$ is continuously improving according to the criterion $\lambda$, the generative layout distribution $\mathbb{Q}_\lambda(\mathbf{X}|G)$ learned by \modelName\ will be as close as possible to the globally optimal layout distribution $\tilde{\mathbb{P}}_\lambda(\mathbf{X}|G)$. The training procedure of \modelName\ is described in Algorithm~\ref{alg:our-gan}.}

\begin{algorithm}[h]
\footnotesize 
    \caption{SmartGD}
    \label{alg:our-gan}
    \SetAlgoLined
    \SetKwInput{KwInput}{Input}                
    \SetKwInput{KwOutput}{Output}       
    \KwInput{Initial good layout example \revise{$\XrG{0}{}$} for each graph $G$ in the dataset; The objective function $\lambda$.}
    
    \For{\upshape training epoch $t$}{
        // Train Discriminator \\
        \For{\upshape $k$ mini-batches in \revise{$\XrG{t-1}{}$} for $\{G_1...G_m\}$}{
            Generate fake layout \revise{$\XfG{t-1}{}$} for $\{G_1...G_m\}$ 
            Update discriminator $\dis$ with gradient ascent
            \vspace{-3pt}
            \begin{equation}\begin{small}\begin{aligned}
                \nabla_{\dis} \frac{1}{m} \sum_{i=1}^m \ln\sigma \left( 
                    \dis\left(\revise{\XrG{t-1}{i}} \right) 
                    - \dis\left(\revise{\XfG{t-1}{i}} \right)
                \right)
            \end{aligned}\end{small}\end{equation}
        }
        // Train Generator \\
        \For{\upshape $k$ minibatches in \revise{$\XrG{t-1}{}$} for $\{G_1...G_m\}$}{
            Update generator $\gen$ with gradient ascent 
            \vspace{-3pt}
            \begin{equation}\begin{small}\begin{aligned}
                \nabla_{\gen} \frac{1}{m} \sum_{i=1}^m \ln \sigma \left( 
                    \dis\left(\revise{\XfG{t-1}{i}} \right) 
                    - \dis\left(\revise{\XrG{t-1}{i}} \right)
                \right) 
            \end{aligned}\end{small}\end{equation}
        }
        // Self-Challenging \\
        \For{\upshape each \revise{$\XrG{t-1}{}$} in the dataset}{
            Generate layout \revise{$\mathbf{X}_{f,G}$} from $\gen$ \\
            Update dataset $\mathbf{X}_r$
            \vspace{-3pt}
            \begin{equation}\begin{small}\begin{aligned}
                \XrG{t}{} \leftarrow 
                \mathrm{argmax}_{
                    \mathbf{X}\in \left\{
                        \revise{\XrG{t-1}{}}, \revise{\mathbf{X}_{f,G}} 
                    \right\}
                } 
                \lambda(\mathbf{X},G)
            \end{aligned}\end{small}\end{equation}
        }
    }
\end{algorithm}

\subsection{Training and Inference}
\label{sec:training-testing}
 
During the training phase (see \autoref{fig:framework}a), the discriminator will take one input layout at a time and output a goodness score. This input layout can be either the layout generated by the generator or a good layout example. For each epoch, the feedback from the discriminator, which is formulated as the adversarial loss in \autoref{equ:our-gan}, is back-propagated to the generator and discriminator alternately. To be more concrete, the weight of the generator remains unchanged while the adversarial loss is back-propagated to the discriminator, and vice versa. Therefore, the generator and the discriminator have trained alternately so that they are able to co-evolve together.

Once the model converges, the generator encodes the learned generative layout distribution conditional on the graph, enabling layout samples for unseen graphs to be drawn from this distribution. In simpler terms, during the inference stage, only the generator is required to draw an unseen graph. This is depicted in \autoref{fig:framework}c. Here, the trained generator accepts the adjacency matrix of an unseen graph as input and outputs the 2-dimensional node positions as the graph layout.
\begin{figure}[htbp!]
    \setlength{\abovecaptionskip}{-1pt}
    \begin{center}
        \includegraphics[width=1\linewidth]{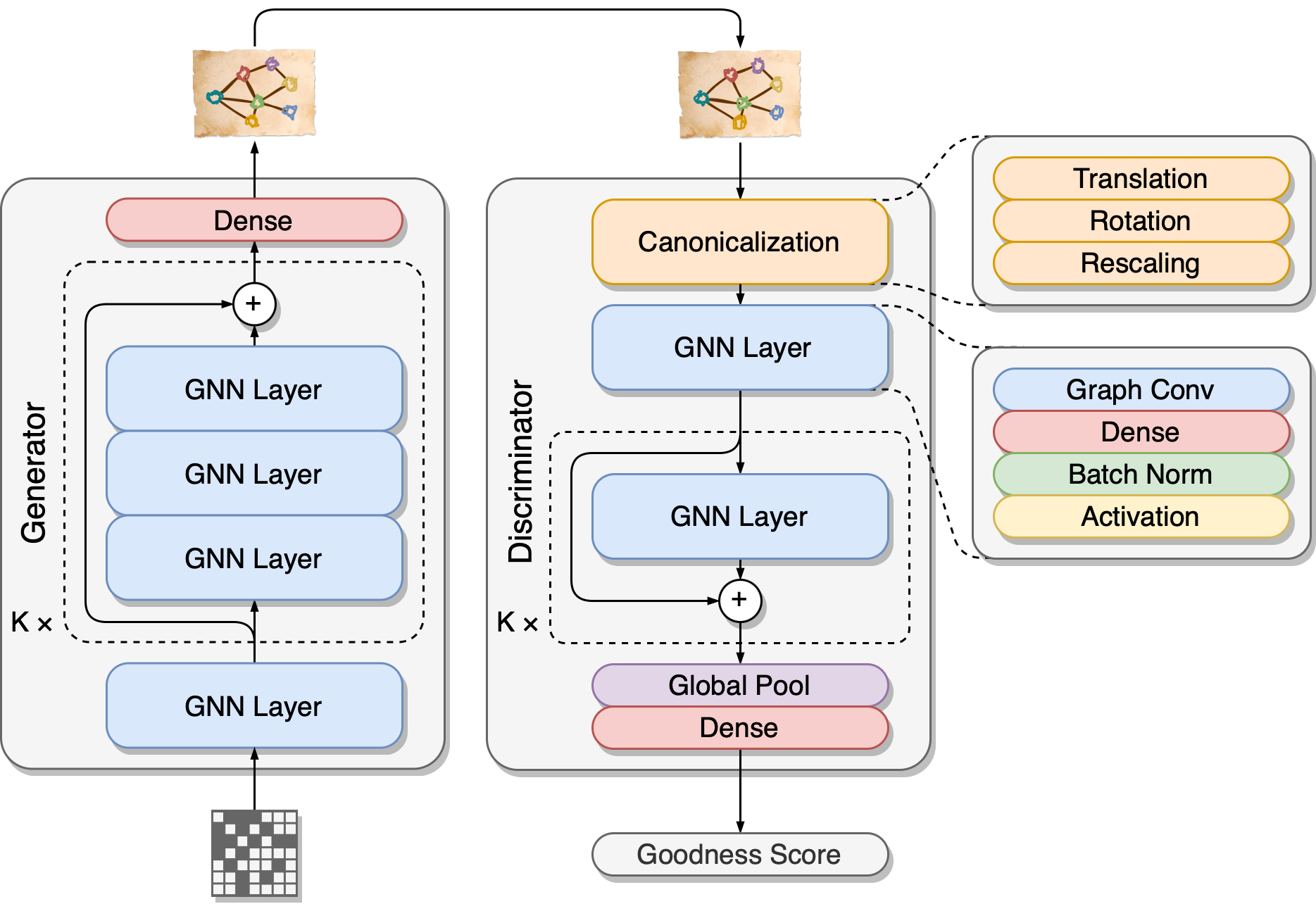}
    \end{center}
    \caption{A unified model architecture of \modelName. \red{K is a hyper-parameter that denotes the number of repetitive GNN blocks.}}
    \label{fig:model-archi}
\end{figure}
\subsection{Model Architecture}
\label{sec:model-archi}

The model architecture is composed of two sub-models: generator and discriminator, as shown in \autoref{fig:model-archi}. The building block of these two sub-models is the GNN layer. Each GNN layer contains a graph convolutional layer (NNConv\cite{nnconv}), a dense (linear) layer, a batch normalization layer, and an activation (LeakyReLU) layer. More specifically, the graph convolutional layer is responsible for generating latent node representations based on graph topology; the dense layer transforms the node representation; batch normalization is adopted to accelerate the convergence by reducing the internal covariant shift; the LeakyReLU activation introduces non-linearity in the model without causing gradient vanishing. 

There are two reasons why we employ graph convolutional layers instead of LSTM layers as DeepDrawing\cite{deep-drawing} proposed. First, it learns a hidden node representation by taking advantage of the message-passing mechanism. For each convolutional layer, the representation of a node is updated based on the aggregated messages passed from its neighbors, which are in turn produced from aggregated messages from their corresponding neighbors in the previous convolutional layer. By stacking multiple convolutional layers, the final node representation will not only contain the local neighborhood information but also capture the global topological structure. Therefore, this allows \modelName\ to draw graphs with \minorrevise{arbitrary topological characteristics}, even if the graph to be drawn \minorrevise{possesses} a completely different topological characteristic than graphs in the training data. Another advantage of the graph convolutional layer is that it does not require \minorrevise{the input graphs to have the same number of nodes.} Each convolutional layer learns a message aggregation function or a kernel function, which is shared across all nodes, to process the messages passed from the neighbors. In this way, the input graph can \minorrevise{have a different number of nodes.} This also endows more flexibility on the general graph drawing framework we proposed.

The node embedding, output by the final GNN layer of the generator, is projected into a 2-dimensional space via a dense layer. In a similar fashion, the initial GNN layer of the discriminator accepts the 2-dimensional node embedding as input. A series of GNN layers within the discriminator then generate the node representation that captures the latent characteristics of the graph layout. Finally, a global mean pooling layer aggregates the representations of all nodes into a singular, graph-level layout embedding. This allows the dense layer to transform the layout embedding into a goodness score.

\subsection{Canonicalization}

As mentioned in Section~\ref{sec:training-testing}, the discriminator will alternately take the layouts generated by the generator and the good layout examples as inputs. However, the generated layout and good layout examples may have inconsistent node position distributions. Additionally, as the model is continuously evolving, the generator is also not guaranteed to produce a stable and consistent node position distribution throughout the training procedure. As a result, an inconsistent or even drastically changing input node position distribution may greatly increase the difficulty for the discriminator to learn. In order to stabilize training by avoiding out-of-distribution inputs, we introduce a canonicalization layer at the beginning of the discriminator as shown in \autoref{fig:model-archi}. It stabilizes the node position distribution by throwing away all the non-essential information for determining the goodness of a layout, including center position, rotation angle, and original numerical scale of node positions. With the canonicalized layouts, the discriminator input is guaranteed to be stable, which will thus speed up convergence, and facilitate generalizability over unseen layout examples.

The canonicalization layer consists of three operations: translation, rotation, and rescaling. \red{Since the gradient needs to be back-propagated over this canonicalization layer during training, each operation needs to be a differentiable function. } First, the translation operation translates each of the node positions in a layout by the same amount in order to \minorrevise{make the input layout be zero-centered}. For each node $i$ in a layout $\mathbf{X}$,
\vspace{-3pt}
\begin{equation}
    \mathbf{X}'_i = \mathbf{X}_i - \frac{1}{N}\sum_{j=1}^N \mathbf{X}_j,
\vspace{-3pt}
\end{equation}
where $\mathbf{X}'_i$ denotes the translated position for node $i$, and $N$ denotes the number of nodes in layout $\mathbf{X}$. Second, the rotation operation rotates the entire layout by its center, such that the first principal component in the layout obtained by PCA is aligned with the x-axis. The direction of the principal components can be found by calculating the eigenvectors of the covariance matrix of the node positions. The inverse of the matrix formed by concatenating the eigenvectors can be used as the rotation matrix. For layout $\mathbf{X}'$,
\vspace{-1pt}
\begin{equation}
    \mathbf{X}'' = \mathbf{X}' \mathrm{eig}^{-1}(\mathrm{cov}(\mathbf{X}')),
\vspace{-1pt}
\end{equation}
where $\mathbf{X}''$ denotes the rotated layout positions, and $\mathrm{cov}(\mathbf{X}')$ represents the covariance matrix of all the node positions in the layout.
Lastly, a rescaling operation is employed to impose a canonical layout scale across different graphs. One way to achieve this is to ensure the scale of node distances in the graph space is consistent with the scale of node distances in the layout space. The discrepancy between graph space and layout space can be measured by stress energy. So we derive an optimal scaling factor by leveraging the equation of stress. For each layout $\mathbf{X}''$,
\begin{equation}
    \mathbf{X}''' = \mathbf{X}'' \cdot \frac{ 
        \sum_{i \neq j} \| \mathbf{X}''_i - \mathbf{X}''_j \| / d_{ij}
    }{
        \sum_{i \neq j} \| \mathbf{X}''_i - \mathbf{X}''_j \|^2 / d_{ij}^2
    },
\end{equation}
where $\mathbf{X}'''$ denotes the rescaled layout positions and $d_{ij}$ represents the graph theoretic distance between node $i$ and $j$.

After performing translation, rotation, and rescaling in sequence, the input layouts of the discriminator will have a canonical representation with the benefit of avoiding the out-of-distribution inputs, facilitating the convergence, and enhancing the generalizability of \modelName\ over unseen graphs.

\section{Evaluation}
\label{sec:evaluation}
\red{In this section, the effectiveness and efficiency of \modelName\ are carefully assessed by comparing against 10 benchmark methods quantitatively and qualitatively. We will describe a detailed experimental study on optimizing differentiable criteria, non-differentiable criteria, and a combination of different aesthetics.}

\subsection{Experimental Setup}
\revise{\modelName\ is implemented with PyTorch\cite{pytorch} and PyTorch Geometric\cite{PyG}.} Every model presented in the following sections is trained on a single Tesla V100 GPU with a memory of 32 GB. For the training configuration, stochastic gradient descent with a minibatch size of 16 graphs is adopted to train \modelName. The optimizer we used is AdamW optimizer with a decay rate of 0.99. The learning rate is initially 0.001 and exponentially decays by a rate of 0.997 for each epoch. Speaking of the model architecture, the generator has 31 GNN layers and the node embedding output from each layer is 8-dimensional; the discriminator has 9 GNN layers and the node embedding output from each layer is 16-dimensional. In total, \modelName\ contains 84,200 parameters in its generator and 1,147,800 parameters in its discriminator. To facilitate faster convergence, the input node embedding of the generator is initialized as a 2-dimensional node position generated by PivotMDS (PMDS)\cite{pmds} with 50 pivots and max iterations of 200 because PMDS can efficiently produce layouts with reasonable quality.

\subsection{Benchmark Algorithms}
\label{sec:benchmark}
To show the effectiveness of \modelName, we compared \modelName\ with 10 benchmark algorithms, including force-directed layouts, energy-based layouts, gradient-based layouts, and deep learning-based layouts. Those 10 benchmarks are widely used layout methods implementing various types of approaches. 

To be precise, spring\cite{spring}, ForceAtlas2 (FA2)\cite{fa2}, and sfdp\cite{sfdp} are three force-directed layout methods aiming at reaching equilibrium by balancing attractive and repulsive forces. Neato\cite{neato} and the method proposed by Kamada and Kawai (KK)\cite{kamada_kawai_1989} are two energy-based layouts in which the stress energy is minimized. $\mathrm{SGD}^2$\cite{zheng-gd2} also attempts to minimize stress by adopting stochastic gradient descent. \minorrevise{The \red{spectral\cite{spectral} layout}} visualizes graphs using the eigenvectors of the graph Laplacian matrix. PivotMDS (PMDS)\cite{pmds} is a sampling-based layout method for efficiently approximating the classical multidimensional scaling layout. $\mathrm{GD}^2$\cite{sgd2} and DeepGD\cite{deepgd} share a common goal of optimizing layouts according to certain differentiable aesthetic criteria but with different approaches. $\mathrm{GD}^2$ adopts stochastic gradient descent to directly optimize the layout, while DeepGD is a GNN-based deep learning model to learn to draw optimal layouts. In order to optimize some non-differentiable aesthetic criteria, including the number of edge crossings, neighborhood preservation, and aspect ratio, a special accommodation is adopted in $\mathrm{GD}^2$ for each of them with the purpose of making them differentiable. However, the authors of DeepGD\cite{deepgd} do not conduct experiments on non-differentiable criteria. Therefore, we will only compare \modelName\ with DeepGD on optimizing stress. Lastly, we also managed to compare \modelName\ against heuristic-based layout methods optimizing edge crossing and crossing angles in the straight-line drawing. However, given that there are only a few existing works\cite{xing-heuristic,Bekos-xangle} and their implementation is not publicly available, we are unable to assess these methods in this paper.

\minorrevise{The implementation of all the benchmarks is from three different sources}, including Graphviz, NetworkX, and the code repositories directly shared by the authors of the papers mentioned above. To evaluate all benchmarks for comparison, the parameter settings we employ are the default ones suggested by Graphviz, NetworkX, and the authors.

\subsection{Datasets}
\subsubsection{\textbf{Graph}}
The graph dataset used in our experiment is Rome graphs\footnote{http://www.graphdrawing.org/data.html}. It contains 11534 undirected graphs containing 10 to 100 nodes. We randomly split the Rome graphs into three sets: a training set with 10000 graphs, a validation set with 534 graphs, and a test set with 1000 graphs. In the following sections, all the \modelName\ models were trained on the training set of Rome graphs and evaluated quantitatively and qualitatively on the test set of Rome graphs.

\subsubsection{\textbf{Good Layout Collections}}
\label{sec:good-layout-collection}
As mentioned in Section~\ref{sec:method}, \modelName\ learns graph drawing by imitating good layout examples. Hence, the quality of the examples is essential to our model performance. If the quality of the good layout collection is better, \modelName\ is more likely to generate a superior layout. \revise{Therefore, for each training graph, we hope to ensure the quality of the layout examples collected is as good as possible.}

In practice, the quality of a layout is usually measured by some commonly agreed aesthetic criteria\cite{purchase1}. Each criterion assesses one aesthetic aspect, and some criteria may even contradict each other\cite{haleem-huamin}. For this reason, it is difficult to find a graph layout that optimizes every aesthetic criterion. Therefore, in our experimental study, \minorrevise{we collected} a separate set of good layout examples for each of the six aesthetic criteria respectively. These six aesthetic criteria include stress, shaped-based metrics\cite{shape-metric}(Shape), the number of edge crossing (Xing), the acute angle formed by a pair of crossing edges (XAngle), a combination of stress and Xing (Stress + Xing), a combination of Stress and Xangle (Stress + XAngle), and a combination of 7 aesthetics (Combined) including stress, Xing, XAngle, the angle formed by two incident edges (IAngle), node occlusion (NodeOcc), uniform edge length (EdgeUni) and the divergence between the graph space and layout space (t-SNE).

To collect the good layout examples for each criterion, a layout for each training graph is generated from 7 existing layout methods in \autoref{table:good-layout}, which are the 10 benchmarks mentioned in Section~\ref{sec:benchmark} except DeepGD\cite{deepgd}, $\mathrm{GD}^2$\cite{sgd2}, $\mathrm{SGD}^2$\cite{zheng-gd2}. The best layout with respect to the criterion among all layouts generated by these methods is then selected as a good layout example for training. If the criterion values of two layouts are tied, the stress is adopted as the tiebreaker. The percentage of every layout method selected into each of the four good layout collections, i.e., generating the winning layouts for each criterion, is presented in \autoref{table:good-layout}. \revise{It is worth mentioning that, even though only the best layout is chosen per graph as the good layout example in our experimental study, it is also possible to employ a sample size greater than one by selecting the top $K$ best layouts.}

\begin{table}[htbp!]
\renewcommand{\arraystretch}{1.1}
\fontsize{7}{7}\selectfont
\setlength{\tabcolsep}{3pt}
\caption{The composition of good layout collections. Each column corresponds to the proportion of a single good layout collection for each criterion. \red{Take the stress as an example, the layouts of 67.80\% training graphs are generated by Neato.}}
\label{table:good-layout}
\centering
\scalebox{0.86}{
\begin{tabular}{c|c|c|c|c|c|c|c}
\bfseries \makecell {Method} & \bfseries Stress & \bfseries Shape & \bfseries \makecell {Xing} & \bfseries \makecell {XAngle}& \bfseries \makecell {Stress+Xing} & \bfseries \makecell {Stress+XAngle} & \bfseries \makecell{7-Aesthetics} \\
\hline
\rule{0pt}{2.2ex}
\bfseries{Neato}\cite{neato}        & 67.80\%  & 11.74\%   & 14.01\%   & 16.5\% & 76.79\% & 81.21\%  & 76.74\% \\
\bfseries{sfdp}\cite{sfdp}         & 0.02\%   & 06.17\%   & 5.47\%    & 6.85\%  & 0.47\% & 0.13\%  & 2.12\% \\
\bfseries{spring}\cite{spring}       & 0.00\%   & 01.61\%   & 3.23\%    & 4.62\%  & 0.27\% &  0.05\% & 7.46\% \\
\bfseries{spectral}\cite{spectral}     & 0.00\%   & 04.83\%   & 1.15\%    & 3.54\%  & 0.00\% & 0.00\%  & 0.00\% \\
\bfseries{KK}\cite{kamada_kawai_1989} & 32.18\%  & 06.50\%   & 8.08\%    & 8.07\%  & 22.04\% &  18.56\% & 13.61\% \\
\bfseries{FA2}\cite{fa2}  & 0.00\%   & 67.41\%   & 67.83\%   & 60.24\% & 0.38\% & 0.03\%  & 0.08\% \\
\bfseries{PMDS}\cite{pmds}         & 0.00\%  & 01.74\%    & 0.35\%    & 0.17\%  & 0.04\% & 0.03\%   & 0.00\% \\
\end{tabular}}
\end{table}

\subsection{Quantitative Evaluation}

\begin{figure}[ht!]
    \setlength{\abovecaptionskip}{-1pt}
    \begin{center}
        \includegraphics[width=\linewidth]{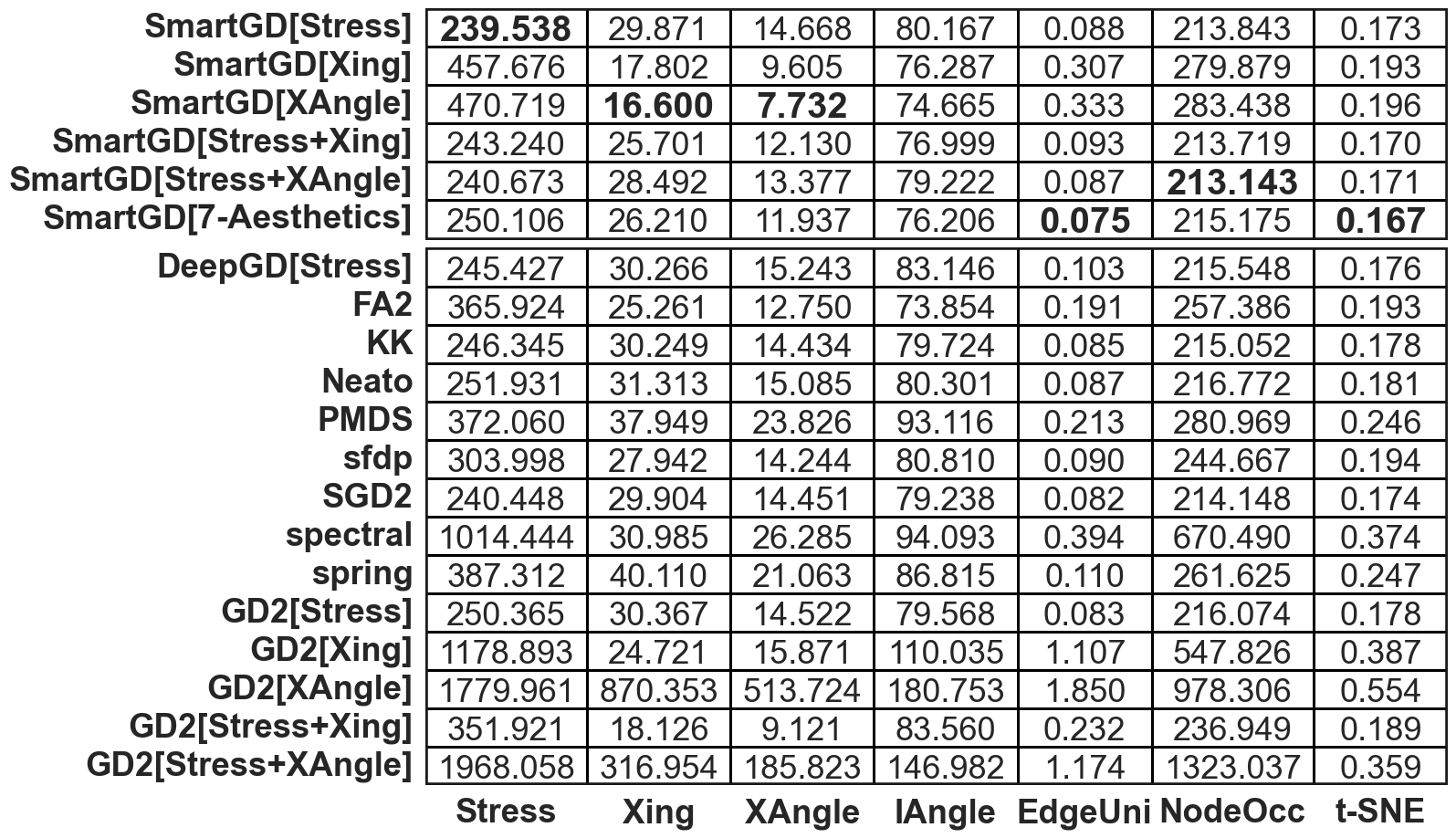}
    \end{center}
    \caption{\revise{The average test metrics of six \modelName\ models (upper) and 14 benchmarks (lower). The optimal value for each criterion among all rows, including both benchmarks and \modelName, is in bold. {\em The detailed equation of each criterion is presented in the Appendix.}}}
    \label{fig:absolute-quant}
\end{figure}

To thoroughly evaluate the effectiveness of optimizing different aesthetic criteria, \red{we trained 7 \modelName\ models with \ganName\ to optimize 4 different aesthetic criteria: stress, shape-based metrics (Shape), edge crossing (Xing), the angle formed by a pair of crossing edges (XAngle), a combination of stress and Xing (Stress+Xing), a combination of stress and XAngle (Stress + XAngle) and a combined criterion which is computed as the weighted average of 7 criteria (Combined). The good layout examples we used for training these 7 \modelName\ models were collected as described in Section~\ref{sec:good-layout-collection}. Similar to DeepGD\cite{deepgd}, the quantitative metric we adopt is symmetric percent change (SPC). For a given graph, SPC measures the relative difference in terms of a given criterion between the \modelName\ layout and the benchmark layout. In this section, the average SPC across the test set is evaluated to assess the relative overall performance of \modelName\ models compared against 10 benchmarks on different criteria of interest. If the average test SPC is negative, this indicates how much the \modelName\ models outperform the benchmark algorithm, and vice versa. \emph{The computation equation of average test SPC is presented in the Appendix.} However, even though the SPC metric allows us to easily understand the relative performance between the two methods, it might convey a misleading message when the criterion value is extremely big or small. Therefore, we also assess \modelName\ models and the benchmarks via another quantitative evaluation metric, which is the average absolute criterion value across the test set (see \autoref{fig:absolute-quant}). }



\begin{figure*}
    \centering
    \includegraphics[width=\linewidth]{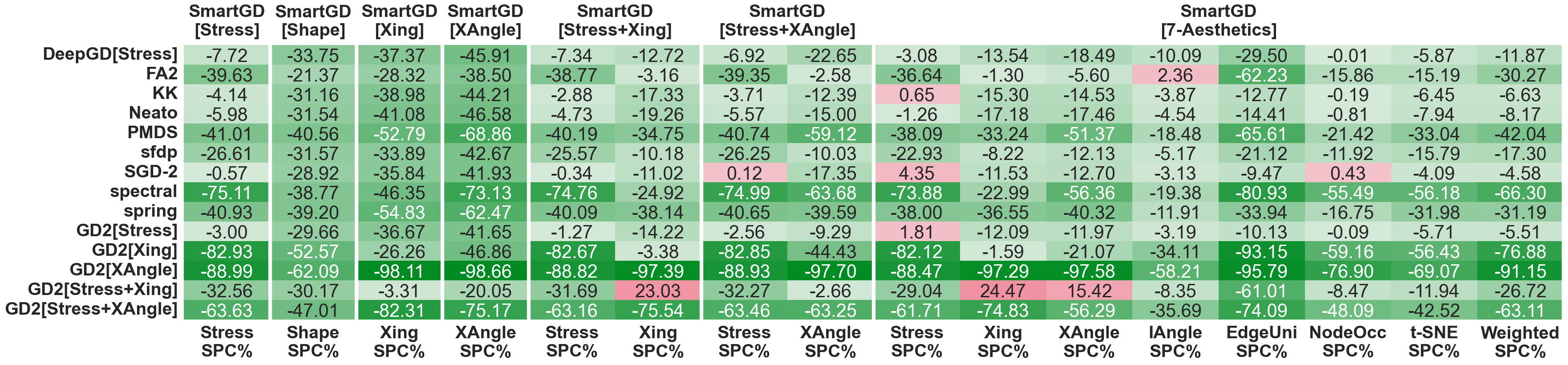}
    \caption{\revise{The mean test SPC of \modelName\ models (column) compared against 14 benchmarks (row). Standard configurations including \ganName, PMDS initialization, and the good layout collection in \autoref{table:good-layout} are used. Green cells indicate where the \modelName\ model (column) outperforms the benchmarks (row), whereas red cells indicate the opposite. The color intensity reflects the magnitude of the difference. For example, the top left cell means that \modelName[Stress] is 7.72\% better in stress than DeepGD[Stress] on average.}}
    \label{fig:spc}    
\end{figure*}

\begin{figure}[ht!]
    \setlength{\abovecaptionskip}{-1pt}
    \begin{center}
        \includegraphics[width=1\linewidth]{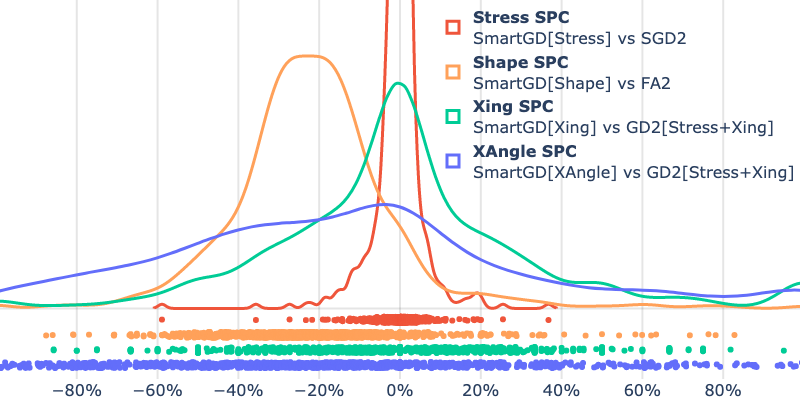}
    \end{center}
    \caption{\revise{The test SPC distribution for four \modelName\ models with respect to their corresponding best-performing benchmark. As shown in the legend, each color corresponds to a unique SPC comparison. The figure contains both the density plot (upper) and the rug plot (lower) for SPC values (x-axis).}}
    \label{fig:single-dist}
\end{figure}

\begin{figure}[ht!]
    \setlength{\abovecaptionskip}{-1pt}
    \begin{center}
        \includegraphics[width=1\linewidth]{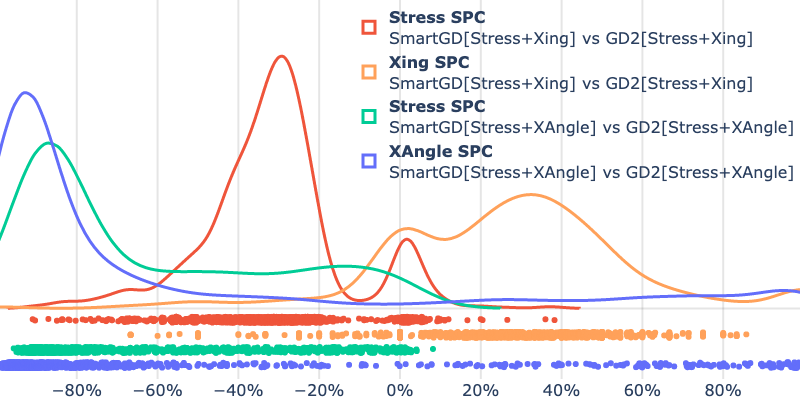}
    \end{center}
    \caption{\revise{The test SPC distribution for two \modelName\ models compared to the corresponding GD2 models trained on Stress+Xing and Stress+XAngle. As shown in the legend, each color corresponds to a unique SPC comparison. The figure contains both the density plot (upper) and the rug plot (lower) for SPC values (x-axis).}}
    \label{fig:2combine-dist}
\end{figure}

\subsubsection{\textbf{Optimizing Differentiable Criteria}}
Stress is a continuous aesthetic criterion that has been shown to be highly correlated to human preference\cite{Tim-user-study}.  Neato\cite{neato}, KK\cite{kamada_kawai_1989} and $\mathrm{SGD}^2$\cite{zheng-gd2} are layout methods that iteratively minimize the stress of a layout. To optimize stress, we train a \modelName\ model using the good layout collection for stress in \autoref{table:good-layout}. As shown in \autoref{fig:spc}, \modelName\ on optimizing stress, abbreviated as \modelName[Stress], achieves negative average stress SPCs computed against all benchmarks. It means that \modelName[Stress] outperforms all benchmarks in terms of stress, among which $\mathrm{SGD}^2$ is the best-performing benchmark. Compared with another deep learning approach optimizing stress, \modelName[Stress] is 7.72\% better than DeepGD[Stress]. Compared with the initial layouts we used, \modelName[Stress] is 41.01\% better than PMDS in terms of stress. In terms of the comparison over the average stress value, \modelName[Stress] also achieves the lowest stress value among all benchmarks (see \autoref{fig:absolute-quant}). Additionally, the distribution of stress SPC for \modelName[Stress] vs. $\mathrm{SGD}^2$ is plotted in \autoref{fig:single-dist}.
In the density plot, the area under the curve to the left of zero SPC almost equals the area under the curve to the right of zero SPC, which indicates that \modelName[Stress] performs on par with $\mathrm{SGD}^2$. In the rug plot, there are more outlier markers to the left of zero SPC. For instance, the leftmost marker shows that \modelName[Stress] performs more than 60\% better than $\mathrm{SGD}^2$ for one of the test graphs.


\subsubsection{\textbf{Optimizing Non-Differentiable Criteria}}
\red{The major advantage of \modelName\ is that it can optimize non-differentiable criteria without the need to manually define differential surrogate functions. Therefore, we conduct a comprehensive evaluation of our effectiveness in optimizing 3 different non-differentiable criteria including edge crossing, angles formed by a pair of crossing edges, and shape-based metrics\cite{shape-metric}. It is worth mentioning that these non-differentiable criteria cannot be directly optimized by gradient-based methods such as $\mathrm{GD}^2$\cite{sgd2} and DeepGD\cite{deepgd}. Hence, $\mathrm{GD}^2$ approximately optimizes edge crossing and crossing angles by reformulating them into differentiable surrogate functions, while DeepGD does not explore the optimization of non-differentiable criteria. However, since there is no effective surrogate function for shape-based metrics yet, DeepGD and $\mathrm{GD}^2$ cannot be applied to optimize shape-based metrics. }
 
\par \textbf{Edge Crossing.} To minimize edge crossing (Xing), we train a \modelName\ model, abbreviated as \modelName[Xing]. From \autoref{fig:spc} and \autoref{fig:absolute-quant}, we can see that \modelName[Xing] can generate layouts with better edge crossing than all benchmarks according to both the average SPC and the average absolute value. Specifically, compared with the best-performing benchmark, GD2[Stress+Xing], \modelName[Xing] is 3.31\% better than GD2[Stress+Xing] on edge crossing. Also, compared with the initial layouts PMDS, \modelName[Xing] is 52.79\% better on edge crossing. The distribution of the test SPC for \modelName[Xing] vs. GD2[Stress+Xing] is plotted in \autoref{fig:single-dist}. 

\par \textbf{Crossing Angle.} To maximize the crossing angle (XAngle), we train a \modelName\ model abbreviated as \modelName[XAngle]. According to the results in \autoref{fig:spc} and \autoref{fig:absolute-quant}, \minorrevise{\modelName[XAngle] outperforms} all benchmarks by at least 20.05\%. Compared with the initial layouts PMDS, \modelName[Xangle] is 68.86\% better on crossing angle. From the rug plot in \autoref{fig:single-dist}, we can observe more red markers to the left of zero SPC, which indicates that \modelName[XAngle] achieves better XAngle than its best-performing benchmark, GD2[Stress+Xing], for more test graphs. 

\par \textbf{Shape-Based Metric.} \revise{Shape-Based metrics\cite{shape-metric} proposed by Eades et al., are designed to measure how similar the shape of the set of node positions is to the original graph.} In our experiments, we train \modelName[Shape] to optimize the shape-faithfulness of graph layouts based on \minorrevise{the relative neighborhood graph (RNG)}. As shown in \autoref{fig:spc}, \modelName[Shape] outperforms all benchmarks by at least 21.37\%. Given that the layout methods optimizing the shape-faithfulness of graph layouts are still much less explored, our promising performance on \minorrevise{the shape-based metric} shows a unique value of \modelName. Besides, the distribution plot in \autoref{fig:single-dist} shows that \modelName[Shape] can obtain better shape-faithfulness than its best-performing benchmark, FA2, \minorrevise{for most of the test graphs. }

\subsubsection{\textbf{Optimizing A Combination of Aesthetics}}
\red{Some research studies have shown that optimizing multiple aesthetic criteria simultaneously is more likely to generate a visually pleasing graph layout\cite{huang-2013}. However, due to the contradictory natures among different aesthetics\cite{haleem-huamin}, greedily optimizing a single aesthetic might compromise some other important aesthetics. Therefore, we also conduct comprehensive experiments on learning to optimize a combination of aesthetics. To the best of our knowledge, $\mathrm{GD}^2$ is the state-of-the-art method in terms of optimizing a combination of criteria. Hence, we conducted experiments on optimizing a weighted average of 1 * stress + 0.2 * Xing and a weighted average of 1 * stress + 0.1 * XAngle, since $\mathrm{GD}^2$ identifies these two pairs of criteria as better pairs or compatible pairs in their paper. The choice of weight factor exactly follows the weight factor settings presented in the $\mathrm{GD}^2$ paper. Besides, in order to show the flexibility of \modelName, we also conducted experiments on training \modelName\ to optimize a combination of 7 different aesthetic criteria. Since $\mathrm{GD}^2$ only conducts experiments on optimizing a pair of criteria simultaneously, we do not present the results of GD2 optimizing more than 2 criteria here. The comparative study against $\mathrm{GD}^2$ \minorrevise{is shown} in \autoref{fig:spc} and \autoref{fig:absolute-quant}. }

\par \textbf{Stress and Xing.} The \modelName\ model optimizing a weighted average of 1 * stress + 0.2 * Xing is abbreviated as \modelName[Stress+Xing]. It is 23.03\% worse in Xing but 31.69\% better in stress than GD2[Stress+Xing]. Given that stress and Xing contradict each other in certain ways, it is difficult to identify a clear winner between \modelName[Stress+Xing] and GD2[Stress+Xing]. However, since stress has a greater weight factor than Xing, \modelName[Stress+Xing] seems to outperform GD2[Stress+Xing] in terms of the weighted average of stress and Xing. The distribution of the test SPC for \modelName[Stress+Xing] vs. GD2[Stress+Xing] is plotted in \autoref{fig:2combine-dist}. From the density plot, the green curve indicates that \modelName[Stress+Xing] is more likely to generate layouts with lower stress than GD2[Stress+Xing]. The rug plot also conveys the same message as the density plot.

\par \textbf{Stress and XAngle.} The \modelName\ model optimizing the weighted average of \minorrevise{1 * Stress + 0.1 * XAngle} is abbreviated as \modelName[Stress+XAngle]. From \autoref{fig:spc}, \modelName[Stress+XAngle] is 63.25\% better in XAngle and 63.46\% better in stress than GD2[Stress+XAngle]. The distribution of the test SPC for \modelName[Stress+XAngle] vs. GD2[Stress+XAngle] is plotted in \autoref{fig:2combine-dist}. From the density plot, the area under the blue curve to the left of zero SPC is larger than the area under the blue curve to the right of zero SPC, which indicates that  \modelName[Stress+XAngle] is more likely to generate layouts with lower stress than GD2[Stress+XAngle].

\par
\begin{figure}[ht!]
    \setlength{\abovecaptionskip}{-1pt}
    \begin{center}
    \includegraphics[width=1\linewidth]{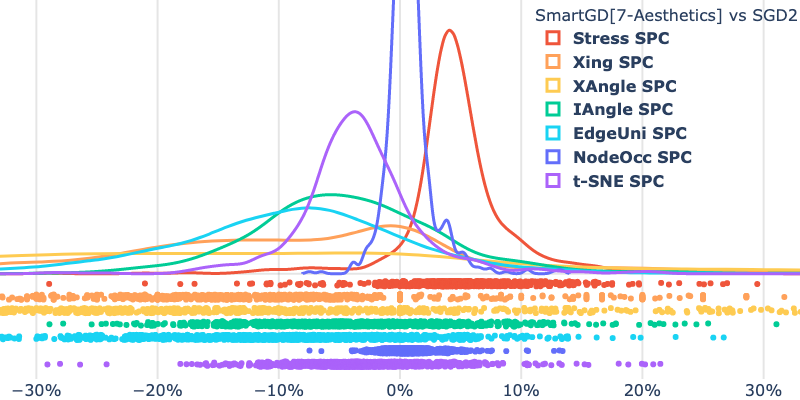}
    \end{center}
    \caption{\revise{The test SPC distribution for \modelName\ optimized on a combination of 7 aesthetics against the best-performing benchmark $\mathrm{SGD}^2$. Each color corresponds to the SPC for a unique \minorrevise{criterion. The} figure contains both the density plot (upper) and the rug plot (lower) for SPC values (x-axis).}}
    \label{fig:combined-dist}
\end{figure}

 \textbf{Combination of 7 Aesthetics.} Empirically, the combined criterion is computed as the weighted average of stress (0.2), Xing (0.05), XAngle (0.1), IAngle (0.1), NodeOcc (0.2), EdgeUni (0.15), t-SNE (0.2), after normalizing each of them according to their corresponding numerical scales. We train a model, \modelName[7-Aesthetics], to optimize this combined criterion. From \autoref{fig:spc}, we can see that \modelName\ on optimizing the combined criterion obtained better layouts compared with all the benchmarks from 7 different aesthetic aspects, according to the last column in \autoref{fig:spc}. According to the results shown in \autoref{fig:absolute-quant}, \modelName[7-Aesthetics] achieves the best EdgeUni and t-SNE compared with all the benchmarks and other \modelName\ models. Additionally, the distribution of test SPC against the best-performing benchmark $\mathrm{SGD}^2$ is shown in \autoref{fig:combined-dist}. The distribution plot of \minorrevise{test SPCs} indicates that \modelName[7-Aesthetics] can consistently produce layouts that have better quality in t-SNE, EdgeUni, IAngle, XAngle, and Xing simultaneously than $\mathrm{SGD}^2$ due to the larger area under the curve to the left of zero SPC in the density plot and more markers to the left of zero SPC in the rug plot. In terms of node occlusion, the layouts generated by \modelName[7-Aesthetics] usually achieve similar performance as $\mathrm{SGD}^2$. For stress, $\mathrm{SGD}^2$ achieves slightly better performance than \modelName[7-Aesthetics]. However, given that \modelName[7-Aesthetics] attempts to optimize 7 different aesthetics simultaneously and some of the aesthetics contradict with minimizing stress, the performance of \modelName[7-Aesthetics] regarding stress is reasonably good.


\begin{figure}[ht!]
    \setlength{\abovecaptionskip}{-1pt}
    \begin{center}
    \includegraphics[width=\linewidth]{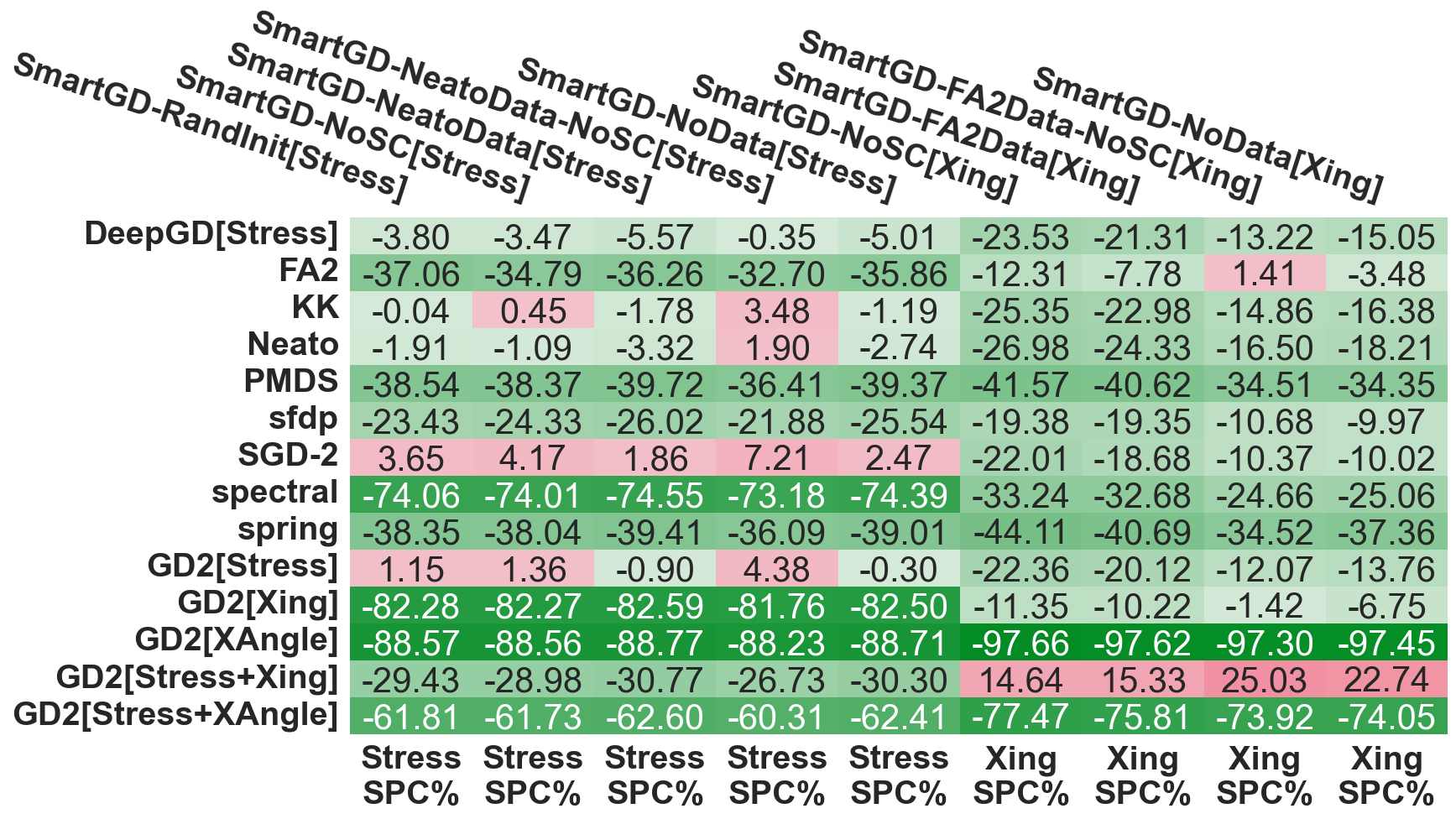}
    \end{center}
    \caption{\revise{The ablation study of \modelName\ on \ganName, PMDS initialization, and good layout collection. This figure follows the same format as Figure~\ref{fig:spc}.}}
    \label{fig:spc-ablation}
\end{figure}

\subsection{Qualitative Evaluation}
We qualitatively evaluate 7 \modelName\ models trained with Rome graphs by presenting the generated layouts of \textit{unseen graphs} in Figure~\ref{fig:vis-result}. Since all the \modelName\ models are only trained with Rome graphs with less than 100 nodes, their performance on drawing large graphs is not always guaranteed to be good. However, it is surprising to see that \modelName\ generates reasonably good layouts for some real-world large graphs from SuiteSparse Matrix Collection\cite{sparse}, even though those large graphs have completely different topological characteristics than Rome graphs in our training data. Therefore, in addition to Rome graphs in the test set, we also present our generated layouts of several large graphs with hundreds to thousands of nodes. For simplicity, only the competitive and representative benchmarks (i.e. $\mathrm{SGD}^2$, PMDS, FA2, DeepGD[Stress], GD2[Stress+Xing]) on quantitative evaluation are selected to be compared qualitatively. To be specific, $\mathrm{SGD}^2$ is the best-performing benchmark on stress and the combined criterion; GD2[Stress+Xing] is the best-performing benchmark on edge crossing and crossing angle; FA2 is a representative of traditional force-directed layouts; DeepGD[Stress] is a representative of deep learning layout methods; PMDS produces the initial layout used in \modelName. Due to the page limit, we only present the qualitative comparison on 8 unseen graphs in Figure~\ref{fig:vis-result}. {\em The qualitative comparison of 95 unseen graphs, including both the Rome graphs and large graphs from the SuiteSparse Matrix Collection, is presented in the Appendix.}

The qualitative comparison in Figure~\ref{fig:vis-result} shows that \modelName\ models optimizing different aesthetics can properly visualize the graphs with various sizes in a visually pleasing and informative way, by satisfying certain aesthetic aspects. It is interesting to observe that \modelName[Xing] and \modelName-NoSC[Xing] tend to bundle edges together to avoid edge crossing. Besides, for visualizing large graphs with GD2[Stress+Xing], we made our best effort to obtain good layouts by running GD2[Stress+Xing] for 2 hours (in addition to the pre-processing time) per graph. However, we \minorrevise{did} not observe any visible improvement after 2 hours of computation. We suspect that directly minimizing edge crossings on large graphs might be a potential weakness of \minorrevise{$\mathrm{GD}^2$\cite{sgd2} since} the loss landscape can be particularly rough for a highly intertwined layout in which an extremely small perturbation in node positions may lead to drastically changing \minorrevise{edge-crossing numbers}.

\subsection{\textbf{Ablation Study}}
\label{sec:ablation}
\red{In this section, we carefully study the effect of different components in the model configuration via a thorough ablation study. These components include \ganName\ and training data from different sources. We present their relative performance with respect to the benchmark algorithms in \autoref{fig:spc-ablation}.}

\par \textbf{\modelName\ without the Self-Challenging Mechanism.} \red{The self-challenging mechanism is an indispensable part of \modelName\ because it allows this GAN-based model to learn a quantitative aesthetic goal. We trained two \modelName\ models, \modelName-NoSC[Stress] and \modelName-NoSC[Xing], with the good layout collection in \autoref{table:good-layout} but without \ganName. Compared with the \modelName[Stress], the stress SPC of \modelName-NoSC[Stress] vs. SGD2 increases by 4.74\% after removing \ganName. Compared with \modelName[Xing], the Xing SPC of \modelName-NoSC[Xing] vs. GD2[Stress+Xing] increases by 17.95\% after removing \ganName. We can see that \ganName\ indeed plays an important role in optimizing the quantitative aesthetic goal. However, it is also interesting to observe that, \modelName-NoSC[Stress] and \modelName-NoSC[Xing] still \minorrevise{outperform most of the benchmarks algorithms}, even without \ganName. One potential reason is that the good layout collection they learn from contains the best layouts selected from 7 different benchmarks, \minorrevise{which allows \modelName-NoSC[Xing] and \modelName-NoSC[Stress] to learn} from the strength of various layouts methods and \minorrevise{outperform each of them.}  In a word, even though they do not explicitly know the target layout preference (i.e. minimizing edge crossing and minimizing stress), they can still generate layouts that align with the inherent preference in the good layout collection.}


\par \textbf{\modelName\ with Training Data from a Single Layout Method.} \red{The quality of good layout examples \modelName\ learns from is very crucial to the model performance. Therefore, we conduct experiments on training \modelName\ with good layouts from a single layout method only. To be specific, \modelName-NeatoData-NoSC[Stress] and \modelName-FA2Data-NoSC[Xing] purely learn from Neato layouts and FA2 layouts but without \ganName. The results in \autoref{fig:spc-ablation} show that \modelName-NeatoData-NoSC[Stress] \minorrevise{achieves} a stress SPC of 1.90\% compared with Neato; \modelName-FA2Data-NoSC[Xing] achieve a Xing SPC of 1.41\% compared with FA2. From this result, we can see that they both achieve comparable performance as the layout methods they learn from, but they can hardly outperform the layout methods they learn from. Furthermore, we also conduct two additional experiments on \modelName\ with \ganName\ but learning from a single layout method, which are \modelName-NeatoData[Stress] and \modelName-FA2Data[Xing]. According to their results in \autoref{fig:spc-ablation}, even though the good layout examples have limited quality, \ganName\ can improve the relative performance of \modelName-NeatoData-NoSC[Stress] w.r.t Neato by 5.22\% and boost the relative performance of \modelName-FA2Data-NoSC[Xing] w.r.t FA2 by 9.19\%. In other words, with the help of \ganName, \modelName\ can break through the shackle of low-quality layout examples it learns from by continuously improving the quality of good layout examples, \minorrevise{and finally outperforms} the initial good layout examples.}


\par \textbf{\modelName\ without Training Data.} \red{We further assess the effectiveness of \modelName\ with \ganName\ but without using any good layout examples. To be more clear, at the first epoch of the training procedure, the initial good examples are generated by the generator itself instead of the layout examples collected in \autoref{table:good-layout}. In this case, \modelName\ solely learns from the layouts generated by itself and utilizes the quantitative criteria as guidance to select good layout examples, without the help of layout examples generated by others. In this experiment, we trained two \modelName\ models to optimize the stress and Xing respectively, but without using any good layout examples. \minorrevise{From \autoref{fig:spc-ablation}}, \modelName-NoData[Stress] outperforms all the benchmarks except $\mathrm{SGD}^2$ and obtains a slightly worse performance on stress than $\mathrm{SGD}^2$ by 1.58\%; \modelName-NoData[Xing] outperforms all benchmarks except GD2[Stress+Xing]. In a word, the performance gap between \modelName\ with and without good layout examples is relatively narrow.  Undoubtedly, \ganName\ endows more flexibility on \modelName\ because \modelName\ can still perform reasonably well even without being provided with initial good layout examples.}

\subsection{Discussion}

In addition to the performance evaluation of \modelName, there are some additional issues that \minorrevise{we would like to} discuss. First, to evaluate the robustness and stability of \modelName, \minorrevise{the 10-fold cross-validation} is performed on \modelName[XAngle] over 10 random train-test splits of Rome graphs. The arithmetic mean of the average XAngle SPC against FA2 is $-35.40\pm1.65\%$, after averaging over 10 folds. Given this small standard deviation across different train-test splits, we can see that the performance of \modelName\ is robust to the potential variation in the training data. 

\red{Furthermore, to demonstrate the effect of using PMDS initial layouts, we trained a \modelName\ model on optimizing stress but with random initial layouts, abbreviated as \modelName-Rand[Stress] in \autoref{fig:spc-ablation}. The result shows that the stress SPC against all benchmarks achieved by \modelName-Rand[Stress] is at most 4.22\% higher than the stress SPC against all benchmarks achieved by the \modelName[Stress] with PMDS initialization. \modelName-Rand[Stress] performs slightly worse than \modelName[Stress] but still obtains a similar performance as GD2[Stress], $\mathrm{SGD}^2$, Neato and KK. This indicates that \modelName\ is not very sensitive to the initial layouts. In addition, we also observed that PMDS initialization could facilitate \modelName\ to converge faster. Specifically, \modelName[Stress] converges around 130 epochs while \modelName-Rand[Stress] converges around 220 epochs.} 

\subsection{Layout Computation Time}
\begin{table}[ht!]
\renewcommand{\arraystretch}{1.1}
\fontsize{8}{8}\selectfont
\setlength{\tabcolsep}{11pt}
\caption{Average layout computation time per graph with 10-100 nodes.}
\label{table:time}
\centering
\scalebox{0.93}{
\begin{tabular}{c|r||c|r}
\bfseries \makecell{Method} & \bfseries \makecell{Time} &
\bfseries \makecell{Method} & \bfseries \makecell{Time}\\
\hline
\rule{0pt}{2.2ex}
\bfseries{\modelName\ on CPU}                       & 0.192s & \bfseries{Neato}\cite{neato}                    & 0.342s \\
\bfseries{\modelName\ on GPU}                       & 0.031s & \bfseries{sfdp}\cite{sfdp}                      & 0.283s \\
\bfseries{DeepGD on CPU}\cite{deepgd}               & 0.274s & \bfseries{PMDS}\cite{pmds}                      & 0.021s \\
\bfseries{DeepGD on GPU}\cite{deepgd}               & 0.058s & \bfseries{SGD2}\cite{zheng-gd2}                         & 0.001s \\
\makecell{\bfseries{GD2[Stress]}\cite{sgd2}}        & 13.190s & \bfseries{spring}\cite{spring}                  & 0.014s \\
\makecell{\bfseries{GD2[Xing]}\cite{sgd2}}          & 142.100s & \bfseries{spectral}\cite{spectral}              & 0.013s \\
\makecell{\bfseries{GD2[XAngle]}\cite{sgd2}}        & 16.750s & \bfseries{KK}\cite{kamada_kawai_1989}           & 0.048s \\
\makecell{\bfseries{GD2[Stress+Xing]}\cite{sgd2}}   & 121.300s & \bfseries{FA2}\cite{fa2}                        & 0.376s \\
\makecell{\bfseries{GD2[Stress+XAngle]}\cite{sgd2}} & 22.250s & & \\
\end{tabular}}
\end{table}

To assess the efficiency of \modelName, the layout computation time is evaluated for all 10 benchmarks and \modelName. Specifically, the computation time we report in \autoref{table:time} is calculated as the average time over 1000 test graphs without including the time for pre-processing and post-processing. Note that the computation time for all graph drawing methods is usually proportional to the graph size. Therefore, the computation time we evaluate can be an approximation of the average drawing time per graph with 10-100 nodes. Given that \modelName\ and DeepGD are deep learning models, they can take advantage of parallelism on GPU so that their computation time on GPU is also evaluated. 

From \autoref{table:time}, spring, PMDS, spectral, KK, $\mathrm{SGD}^2$, \modelName\ on GPU and DeepGD on GPU are the first-tier algorithms regarding efficiency because they are significantly faster than others. $\mathrm{GD}^2$ with different criteria is less efficient than others, even though $\mathrm{GD}^2$ tends to be the best-performing benchmark on edge crossing and the crossing angle. Speaking of our training time, it takes 180-360 seconds on average per epoch and usually converges after around 100-300 epochs, depending on the different criteria of interest.

\newcommand{\imgcell}[1]{\raisebox{-0.4\totalheight}{\adjustbox{height=31px, trim={0.08\width} {0.08\height} {0.08\width} {0.08\height},clip}{\includegraphics[]{#1}}}}

\begin{table*}[ht!]
\setlength{\tabcolsep}{0pt}
\renewcommand{\arraystretch}{0}
\fontsize{6}{6}\selectfont
\centering
\begin{tabular}{ c|ccccc|ccccccc }
    \bfseries{\thead{Graph}} & \multicolumn{5}{c|}{\thead{Benchmark Methods}} & \multicolumn{6}{c}{\thead{SmartGD}}\\
    & \bfseries{SGD2}
    & \bfseries{PMDS}
    & \bfseries{FA2}
    & \bfseries{DeepGD}
    & \makecell{\bfseries GD2\\\relax[Stress+Xing]}
    & \makecell{\bfseries SmartGD\\\relax[Stress]}
    & \makecell{\bfseries SmartGD\\\relax[Xing]}
    & \makecell{\bfseries SmartGD\\\relax[Shape]}
    & \makecell{\bfseries SmartGD\\\relax[XAngle]}
    & \makecell{\bfseries SmartGD\\\relax[Stress+Xing]}
    & \makecell{\bfseries SmartGD\\\relax[Stress+XAngle]}
    & \makecell{\bfseries SmartGD\\\relax[7-Aesthetics]}
    \rule[-1ex]{0pt}{0ex} \\ \hline

\makecell{\bfseries grafo2968.32\\N = 32\\M = 41} &
\imgcell{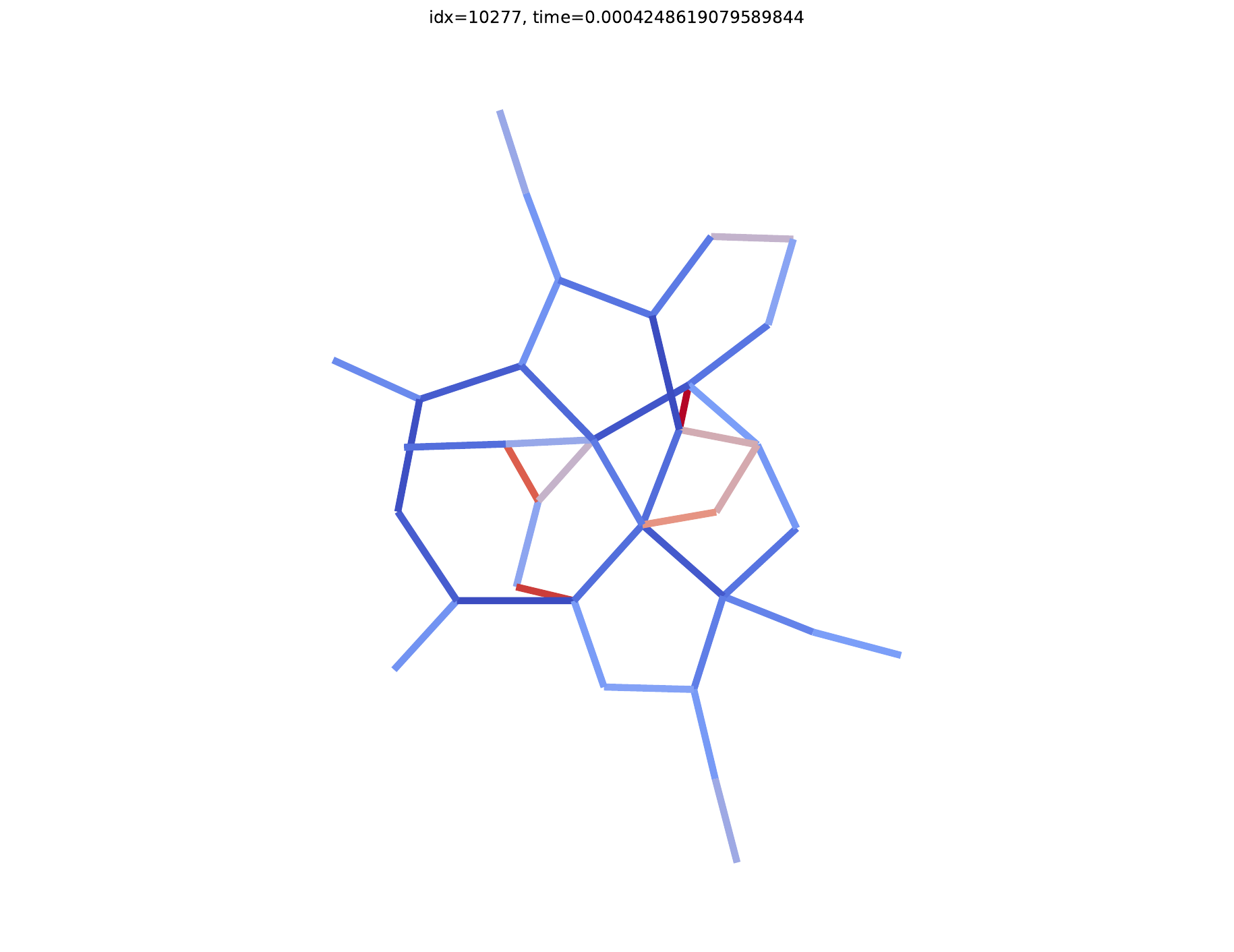} &
\imgcell{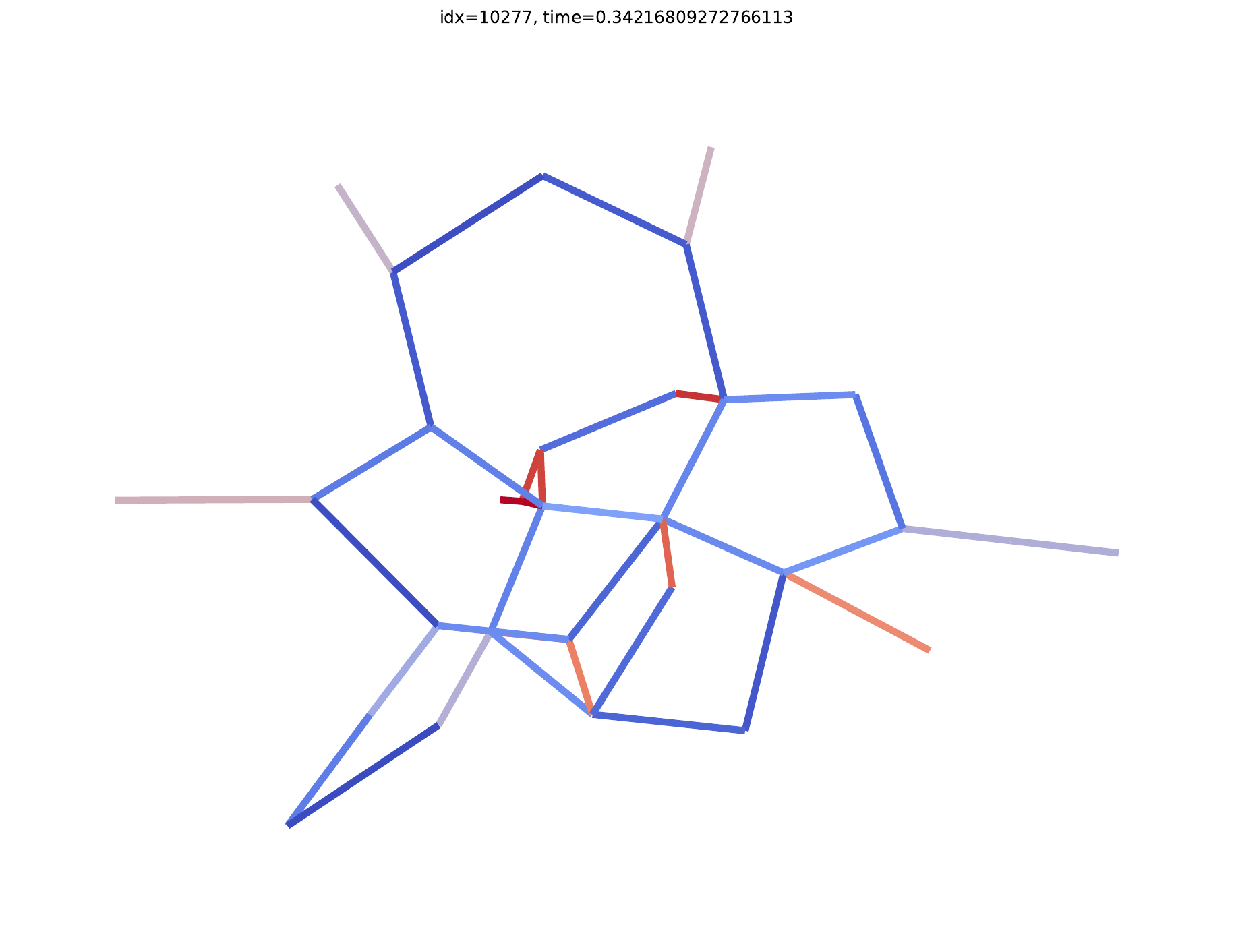} &
\imgcell{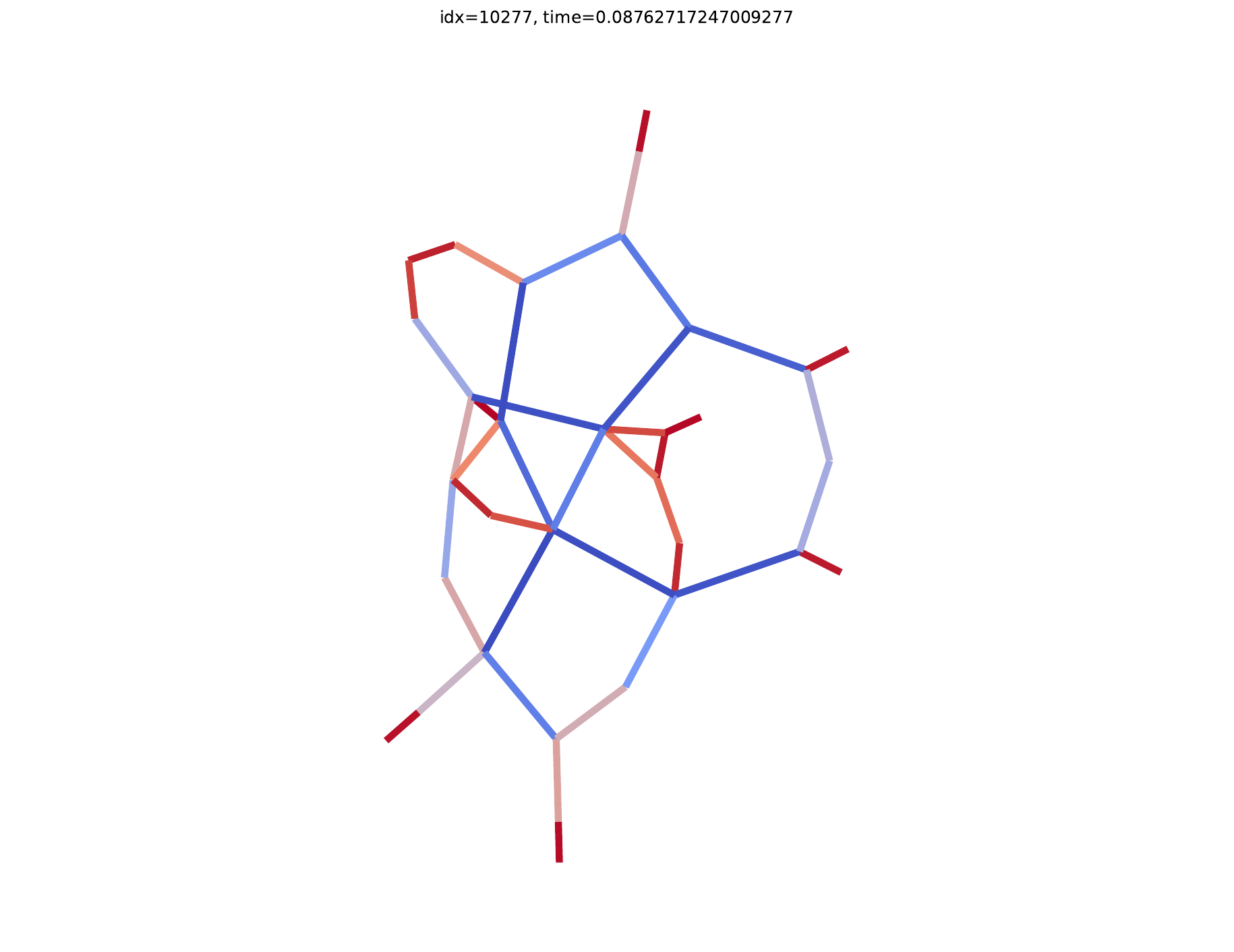} &
\imgcell{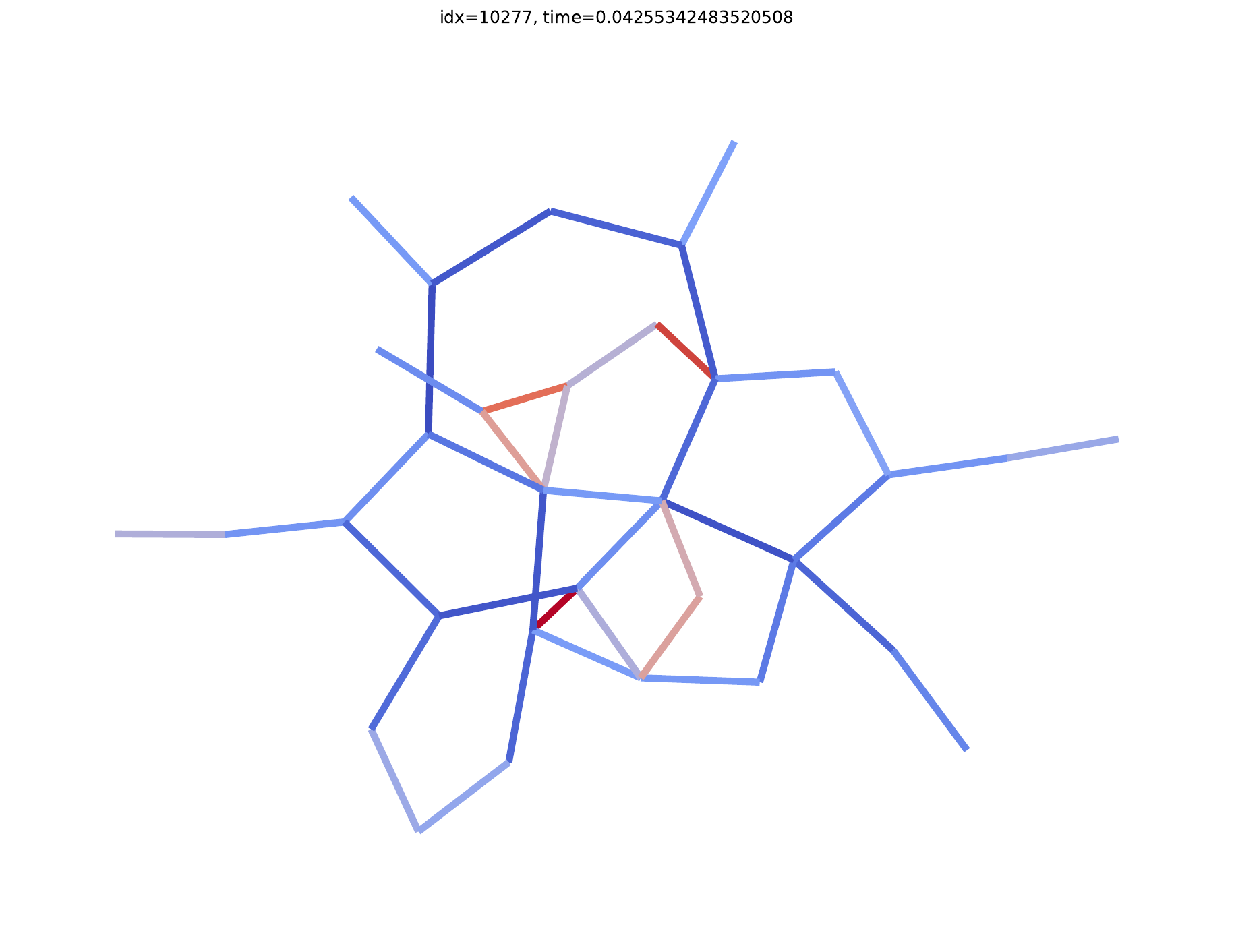} &
\imgcell{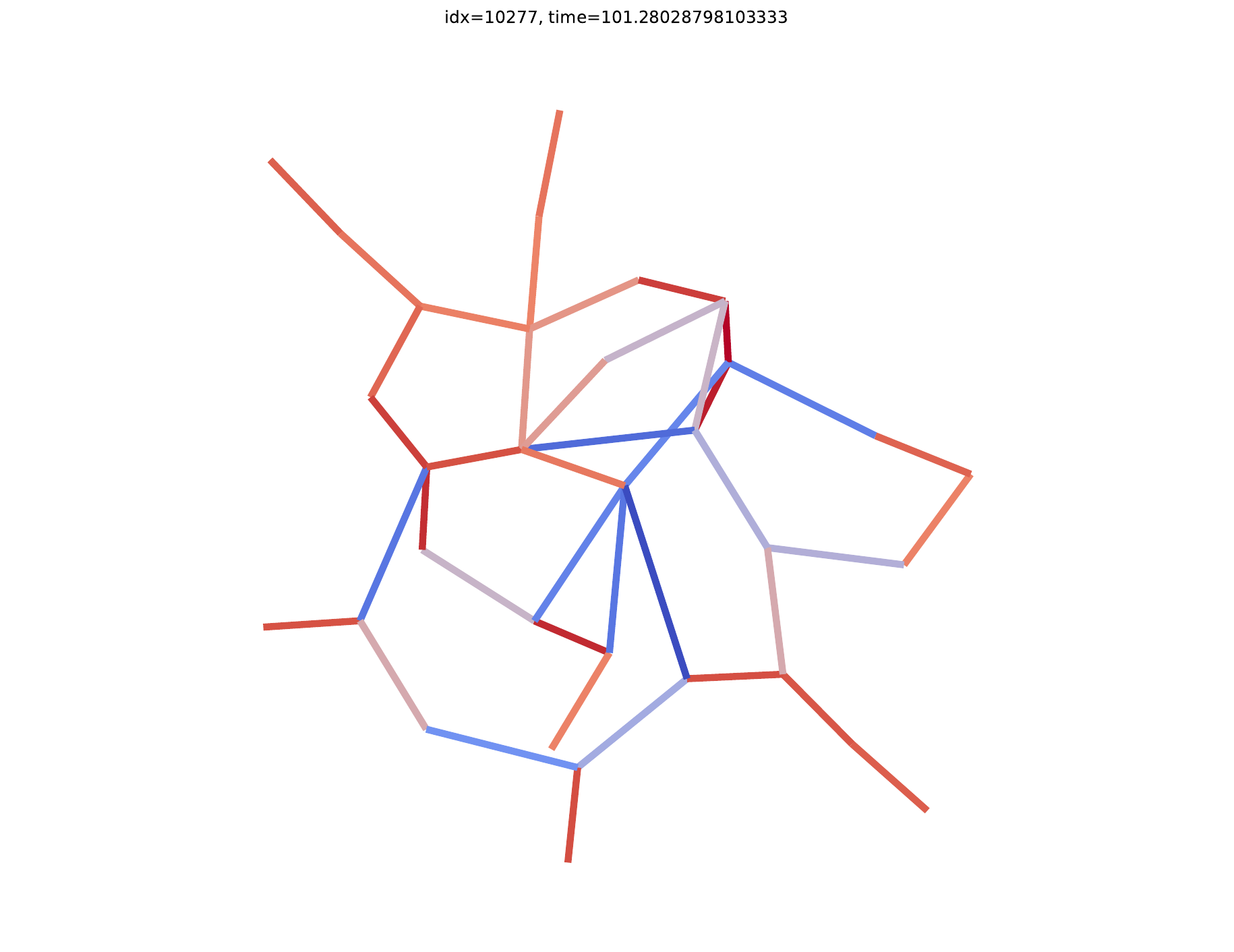} &
\imgcell{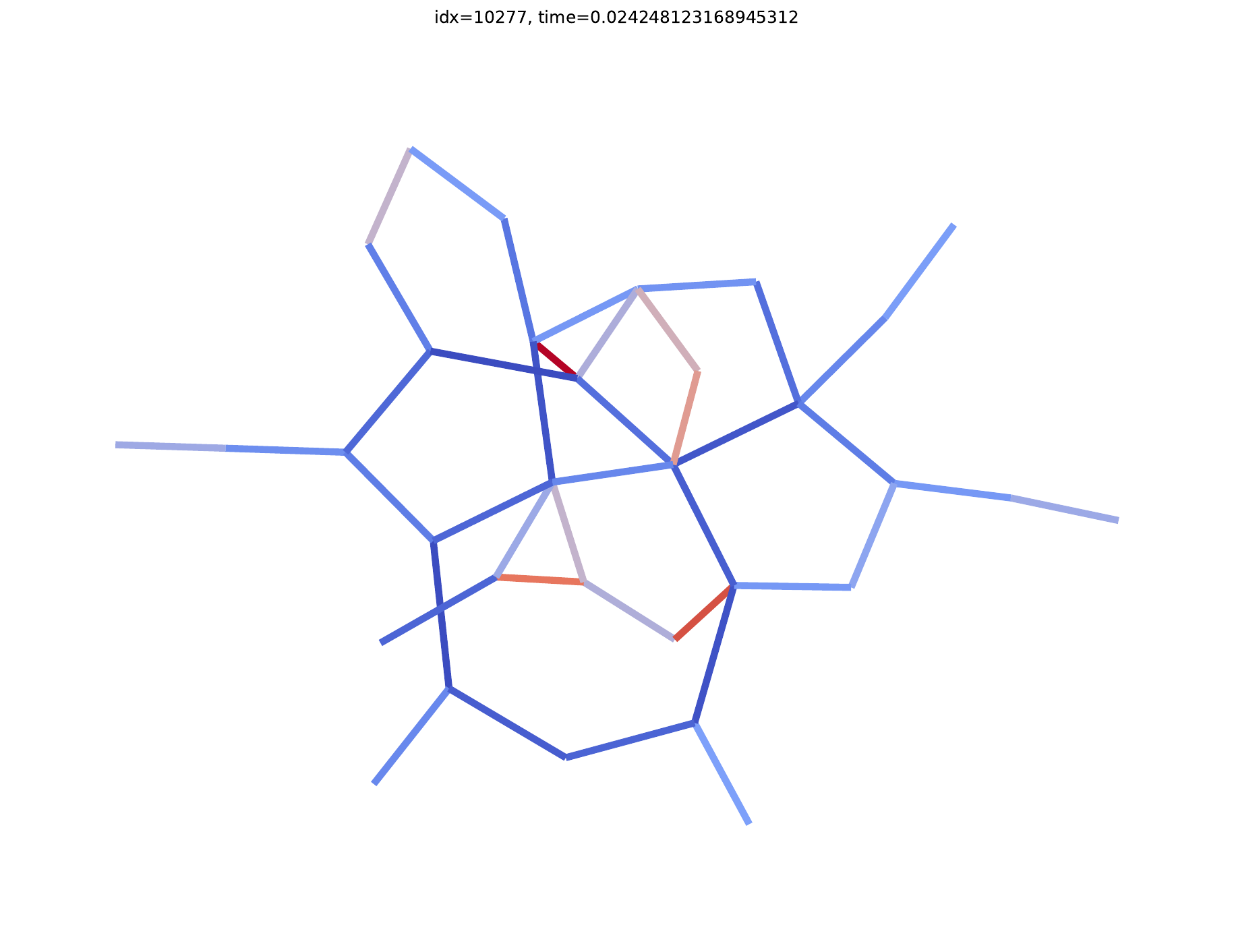} &
\imgcell{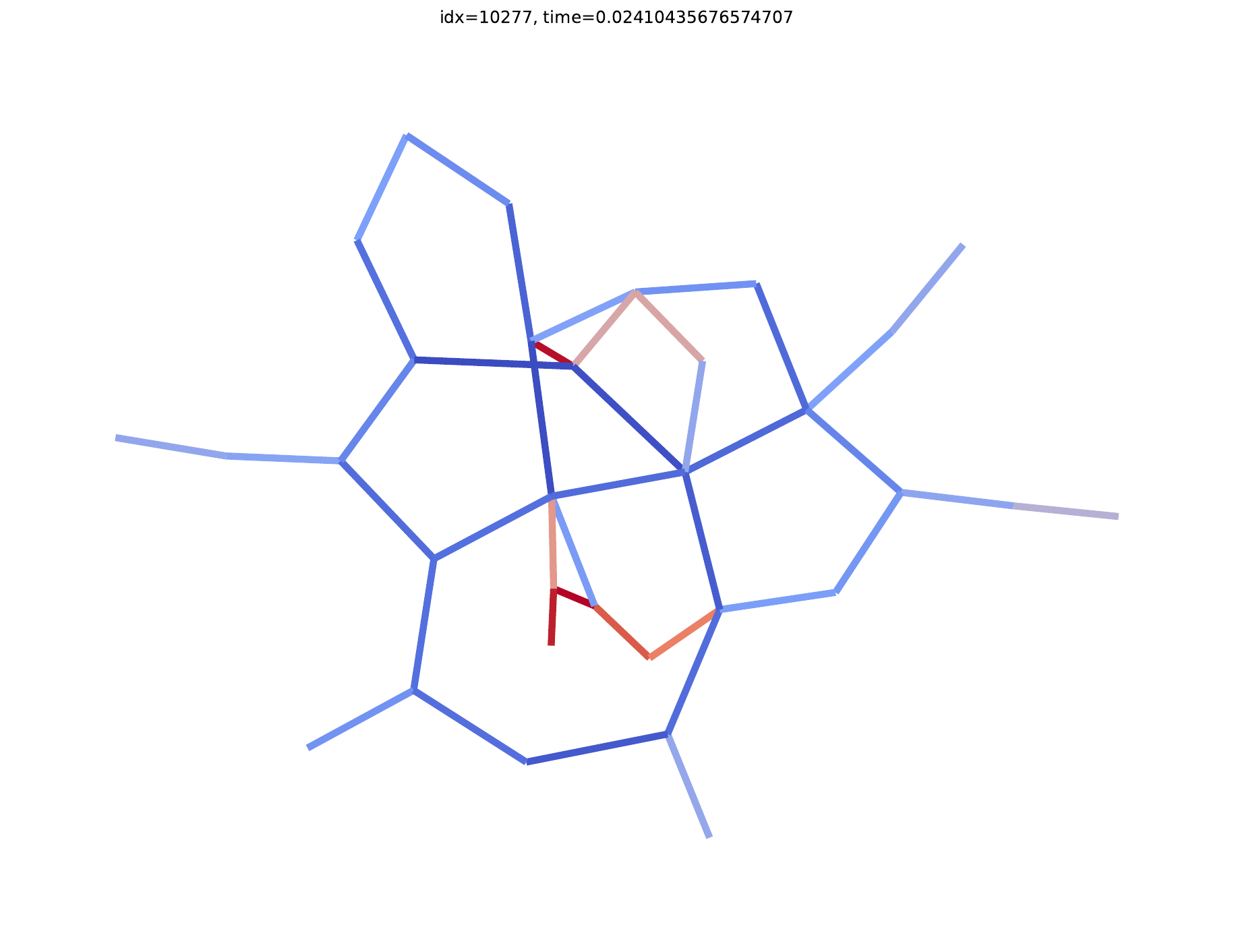} &
\imgcell{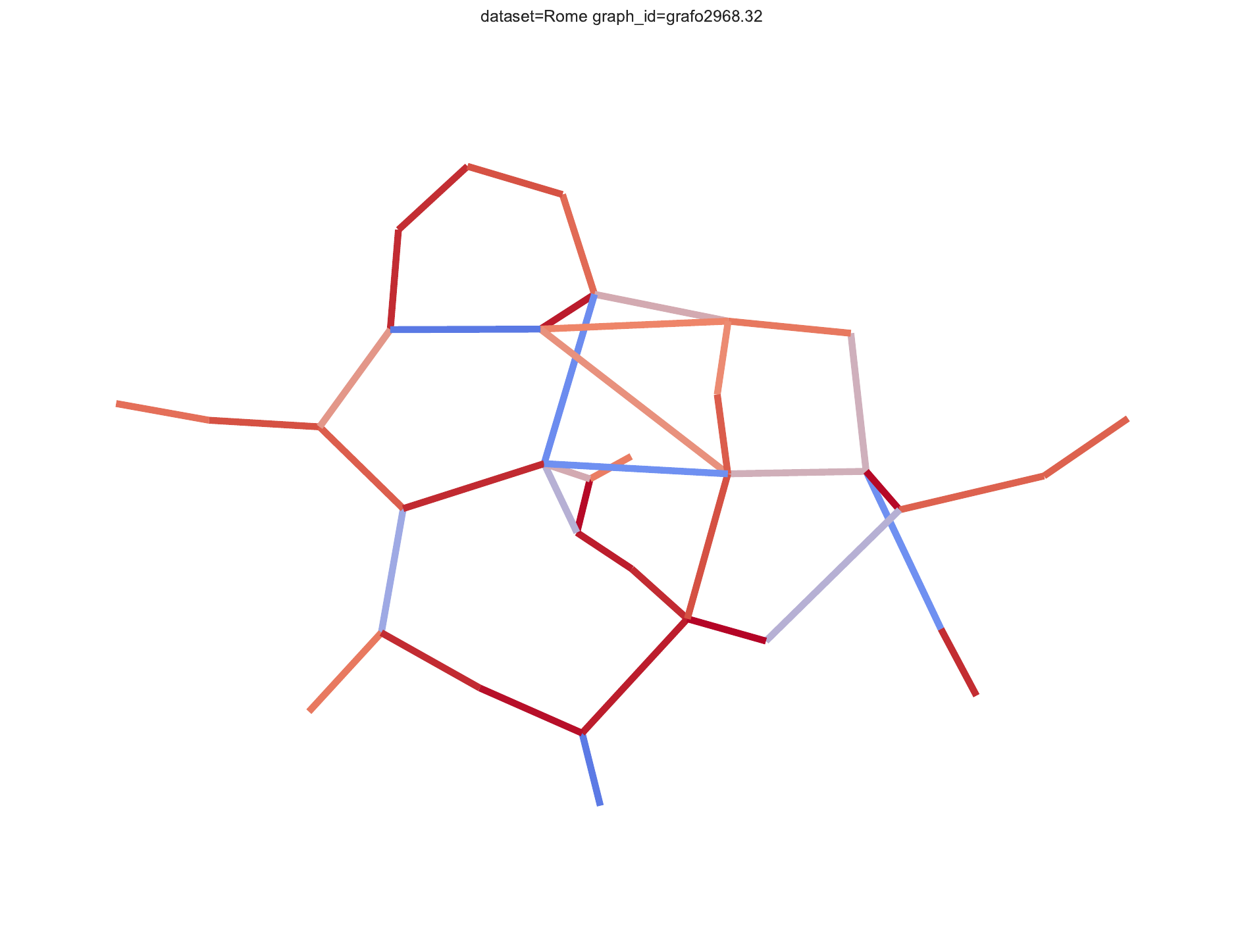} &
\imgcell{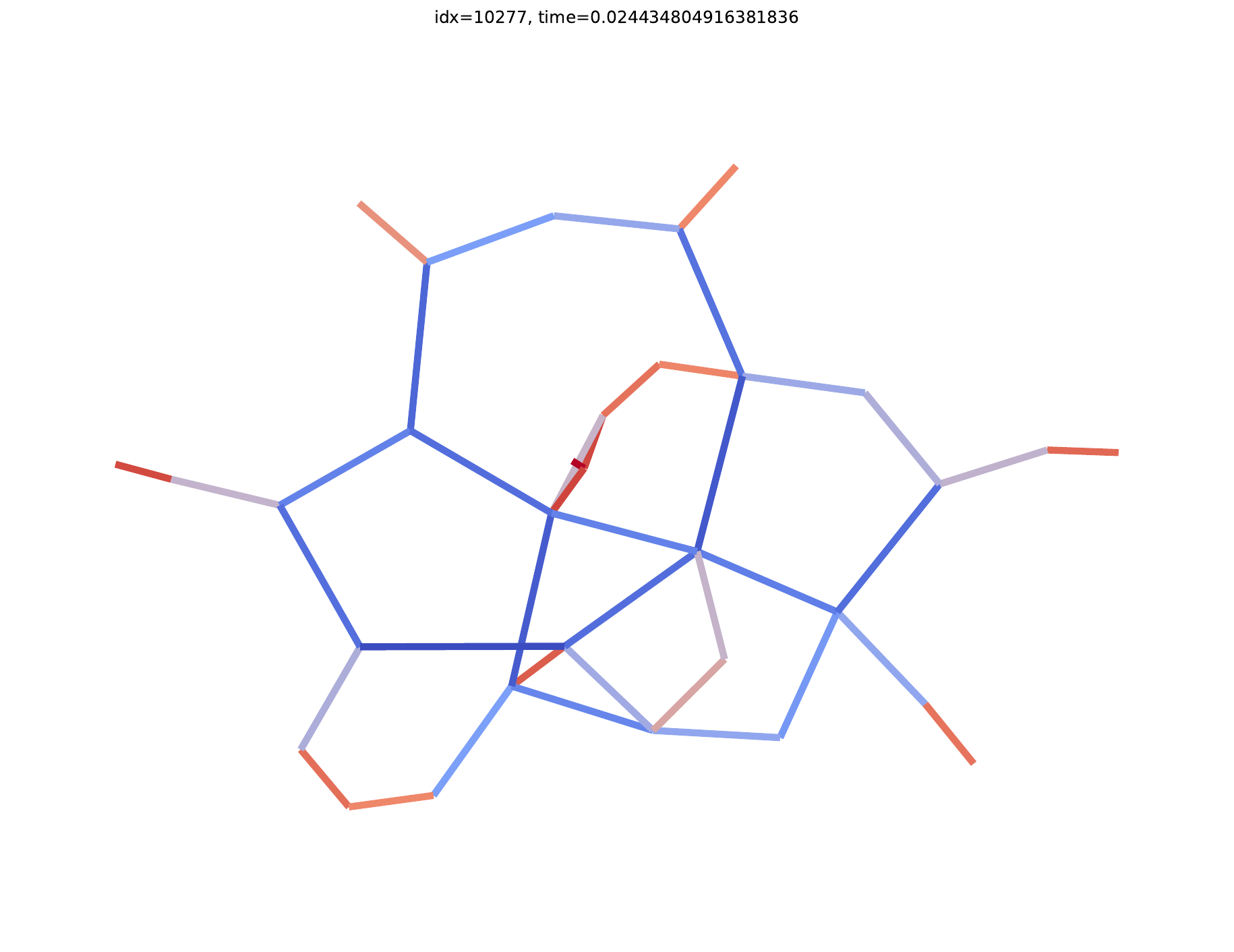} &
\imgcell{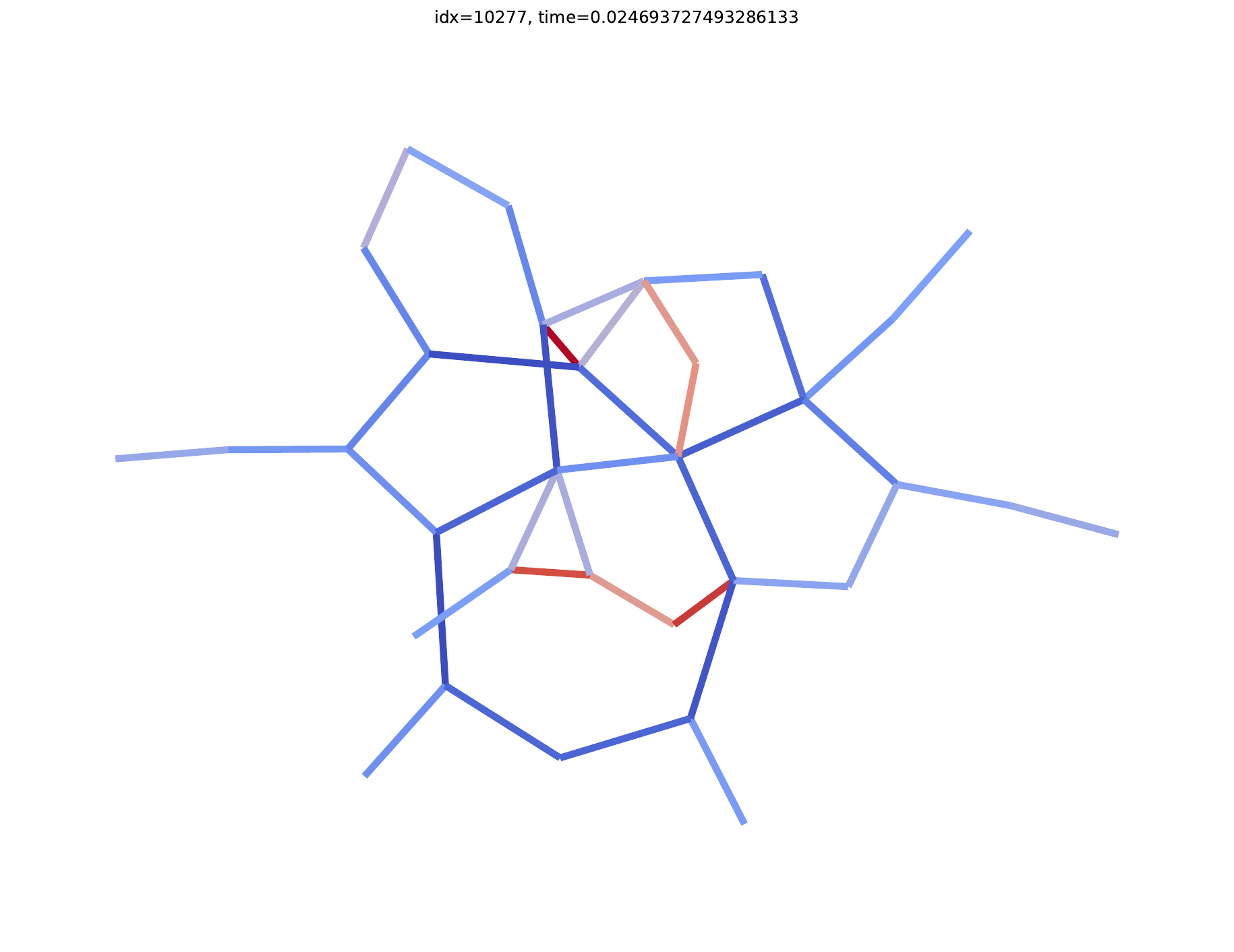} &
\imgcell{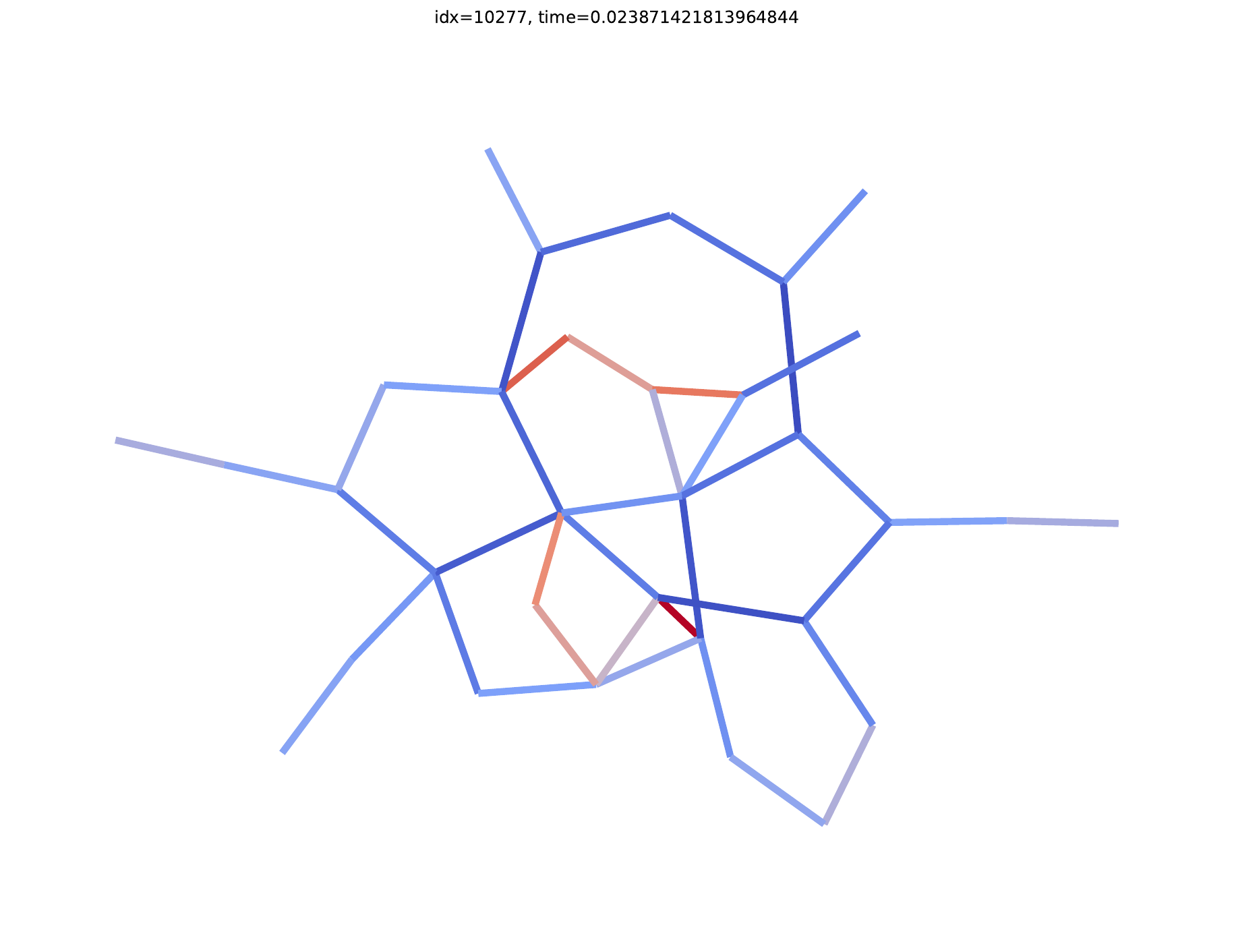} &
\imgcell{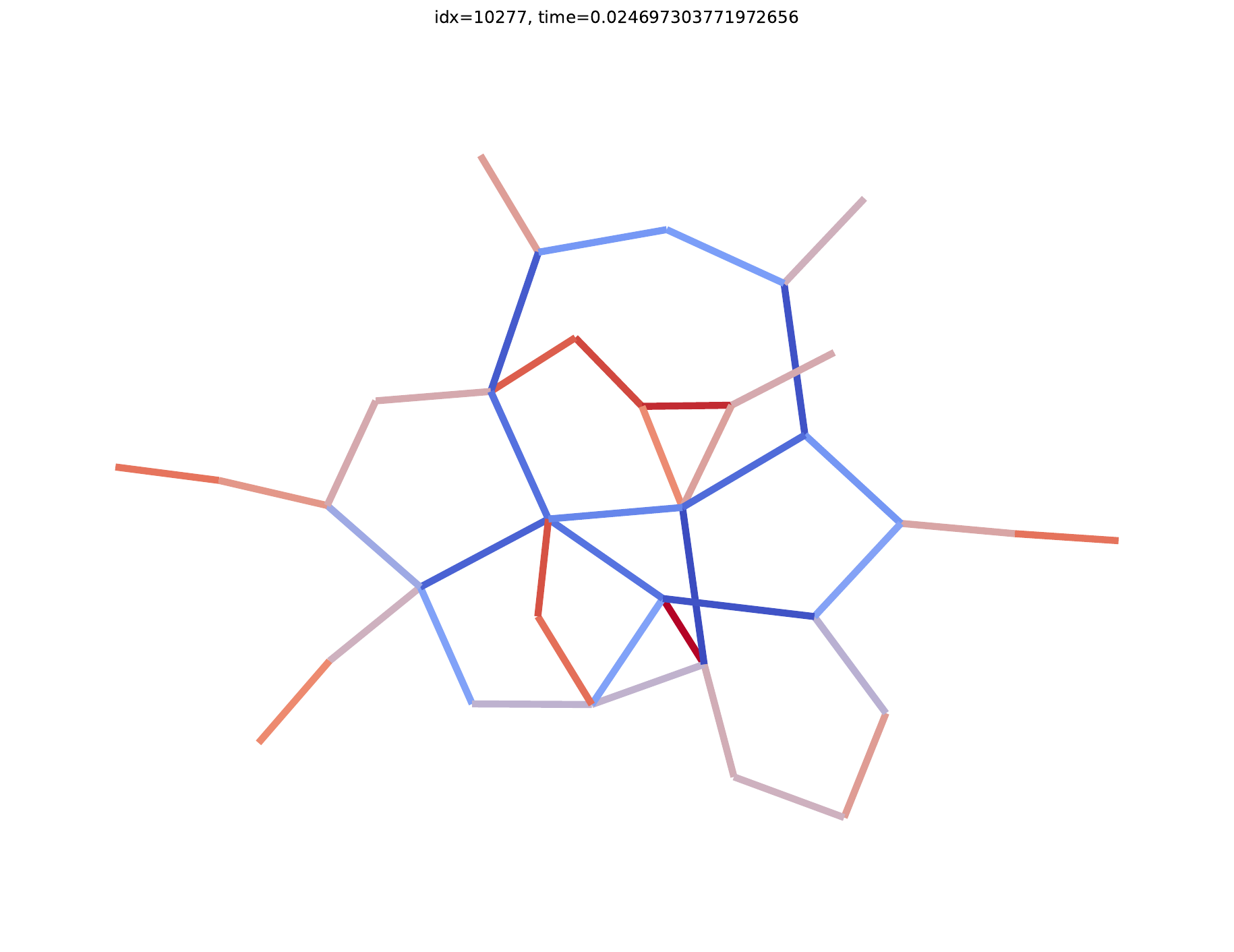} \\

&
t = 0.00s &
t = 0.34s &
t = 0.09s &
t = 0.04s &
t = 101.28s &
t = 0.02s &
t = 0.02s &
t = 0.02s &
t = 0.02s &
t = 0.02s &
t = 0.02s &
t = 0.02s \\

\makecell{\bfseries grafo8900.94\\N = 94\\M = 126} &
\imgcell{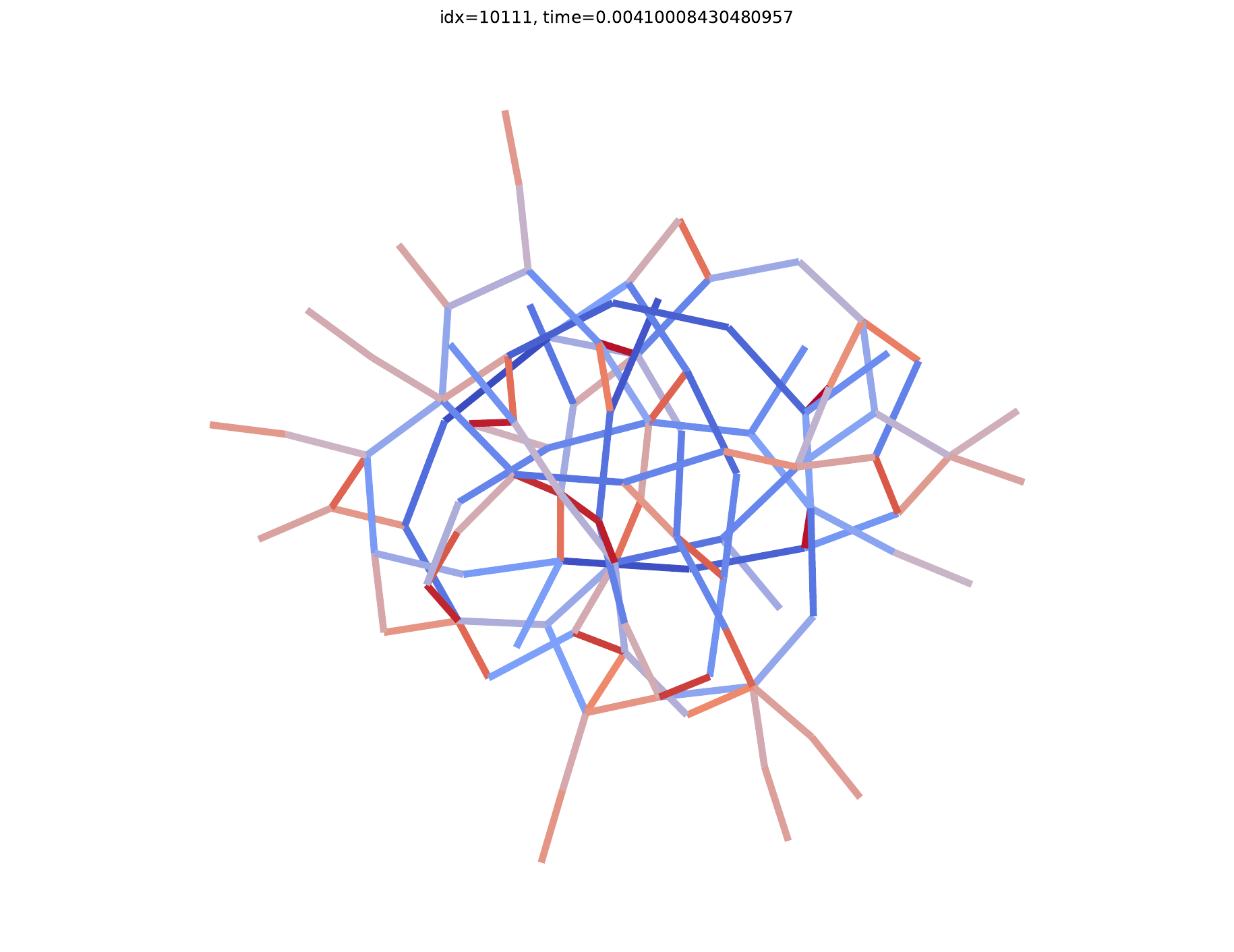} &
\imgcell{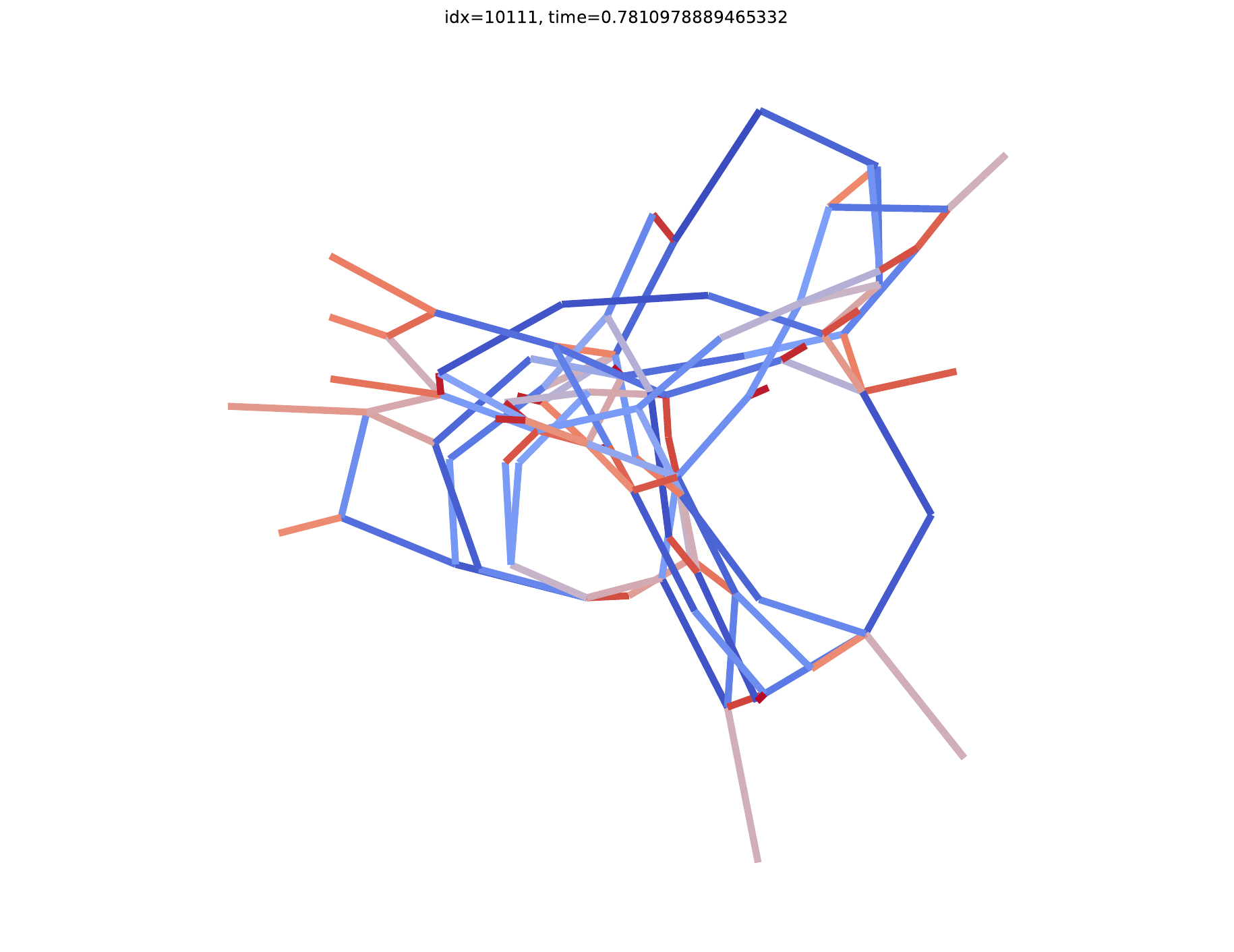} &
\imgcell{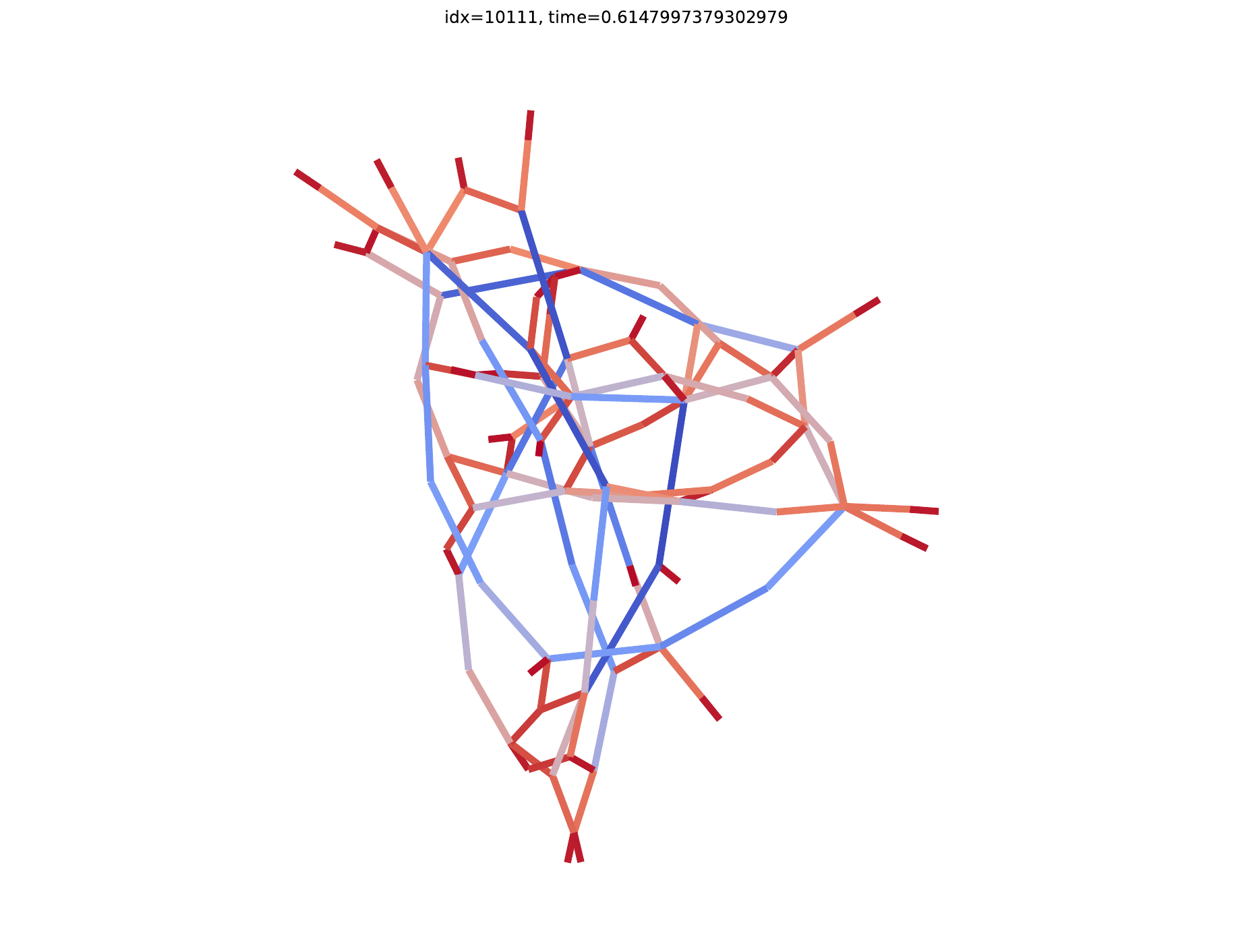} &
\imgcell{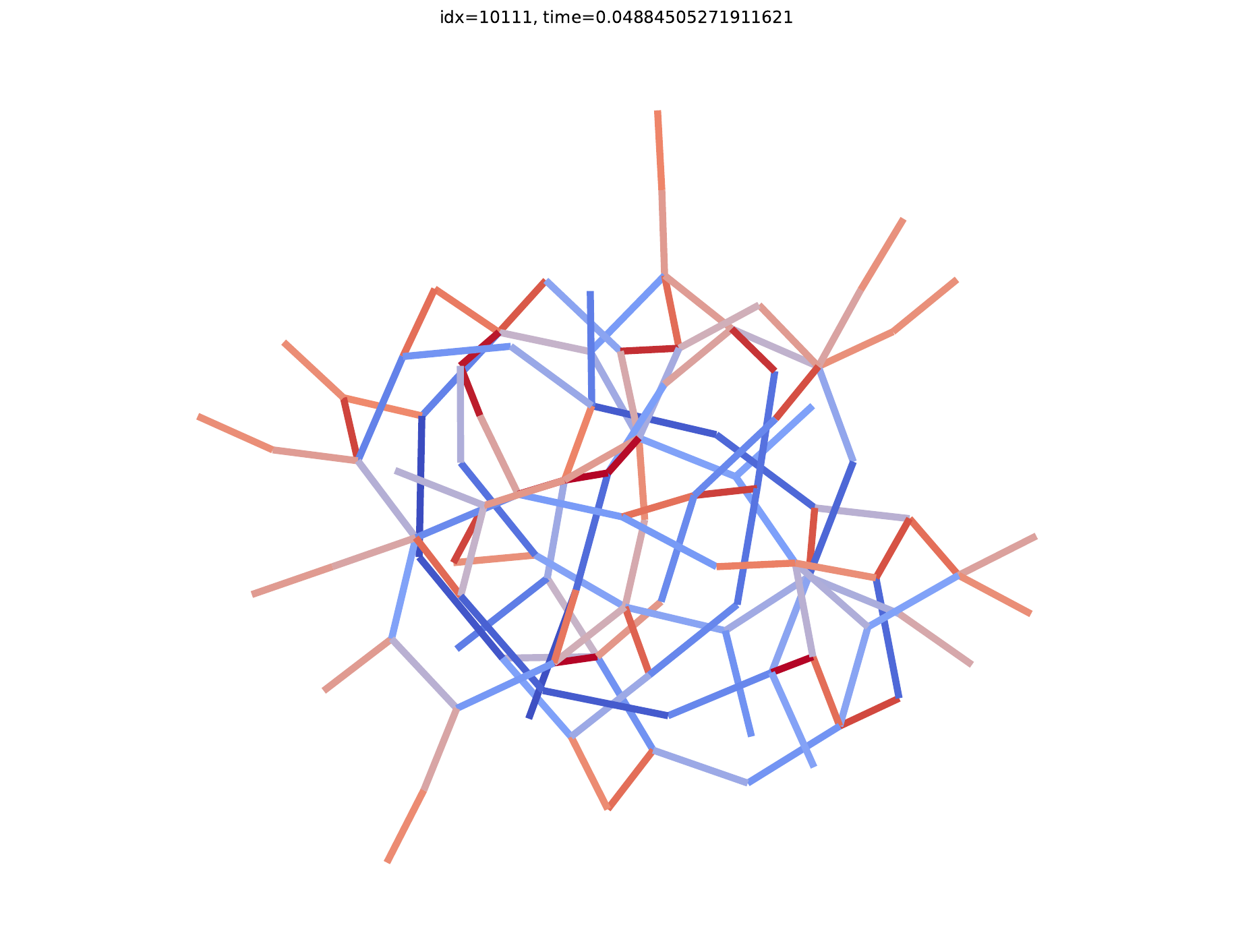} &
\imgcell{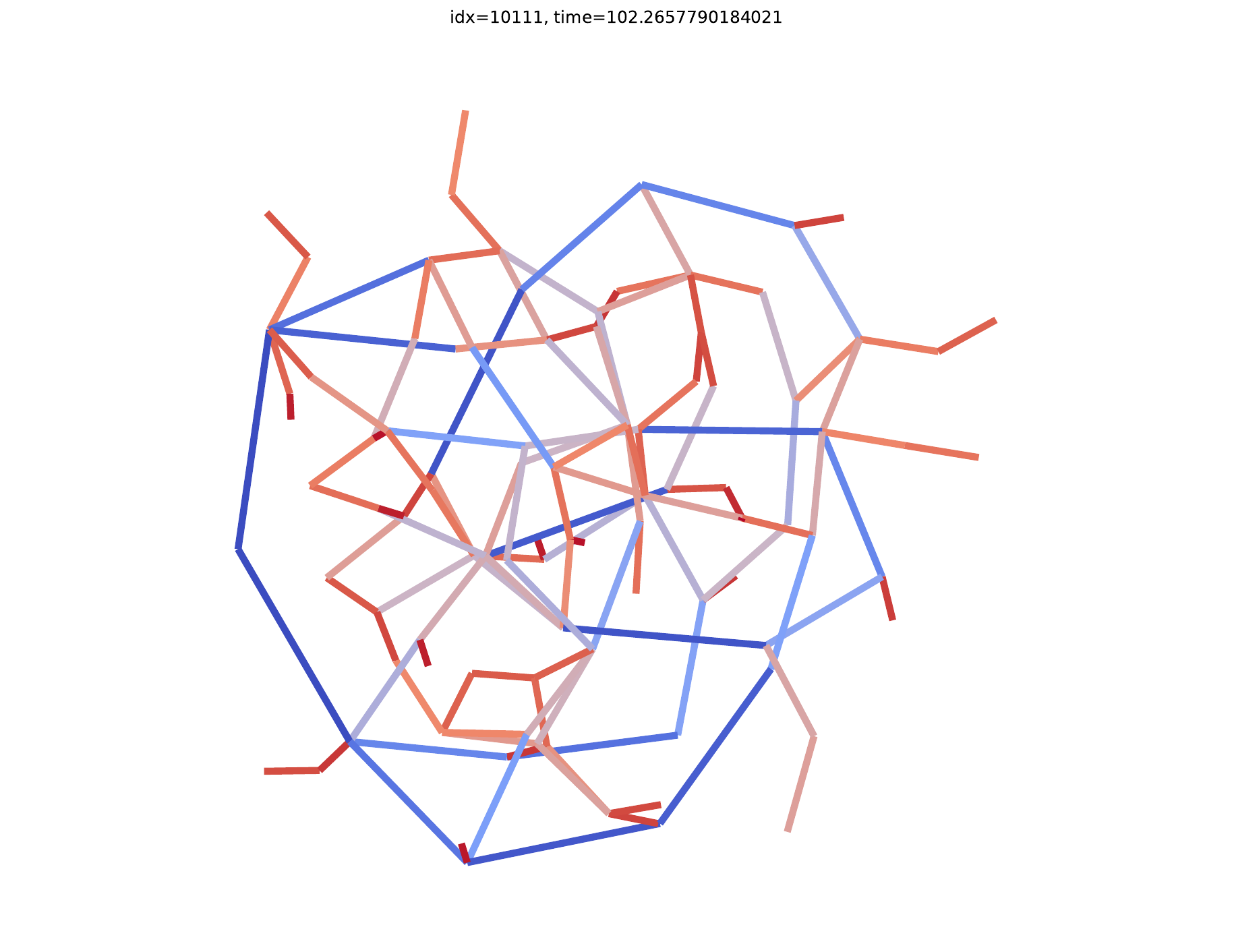} &
\imgcell{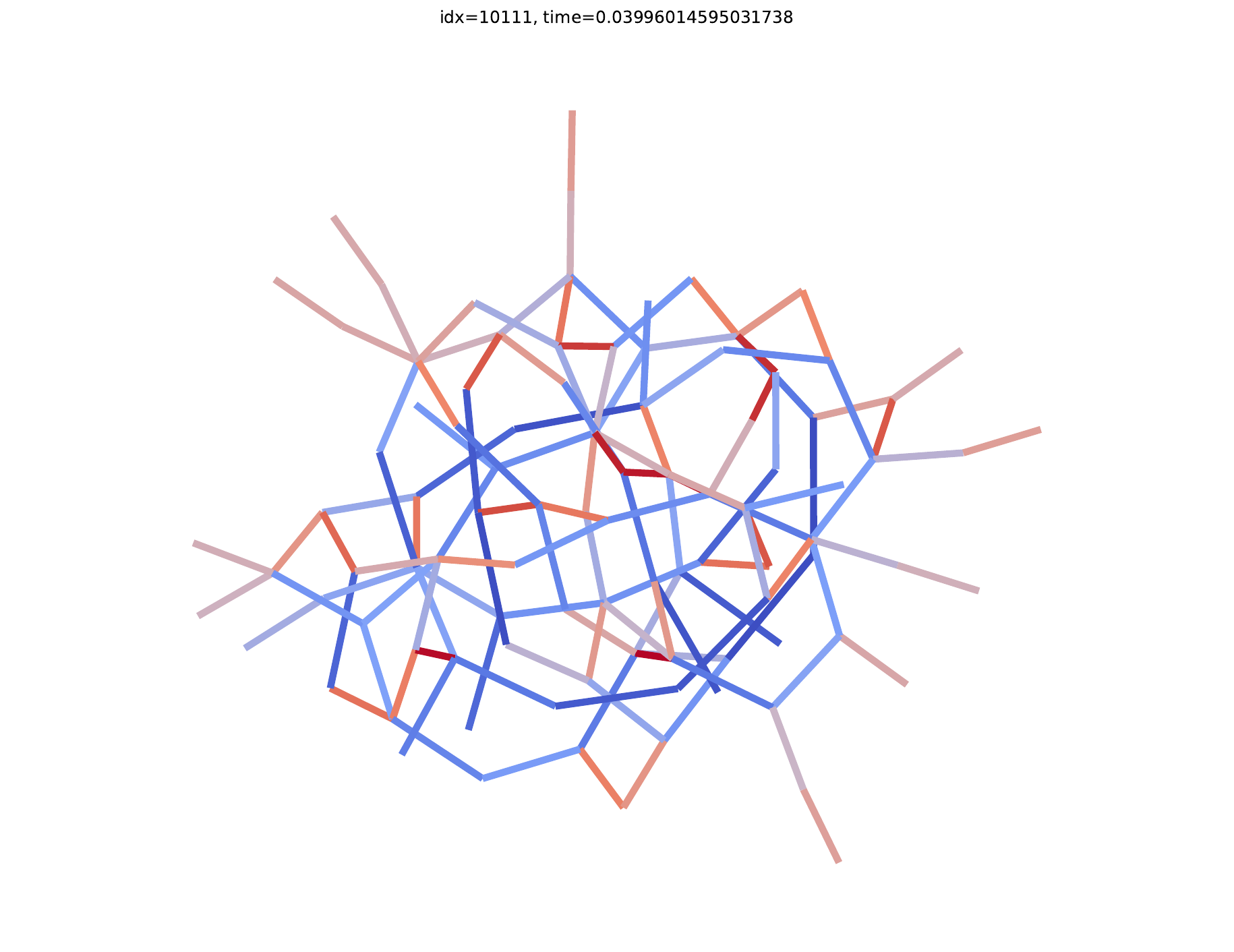} &
\imgcell{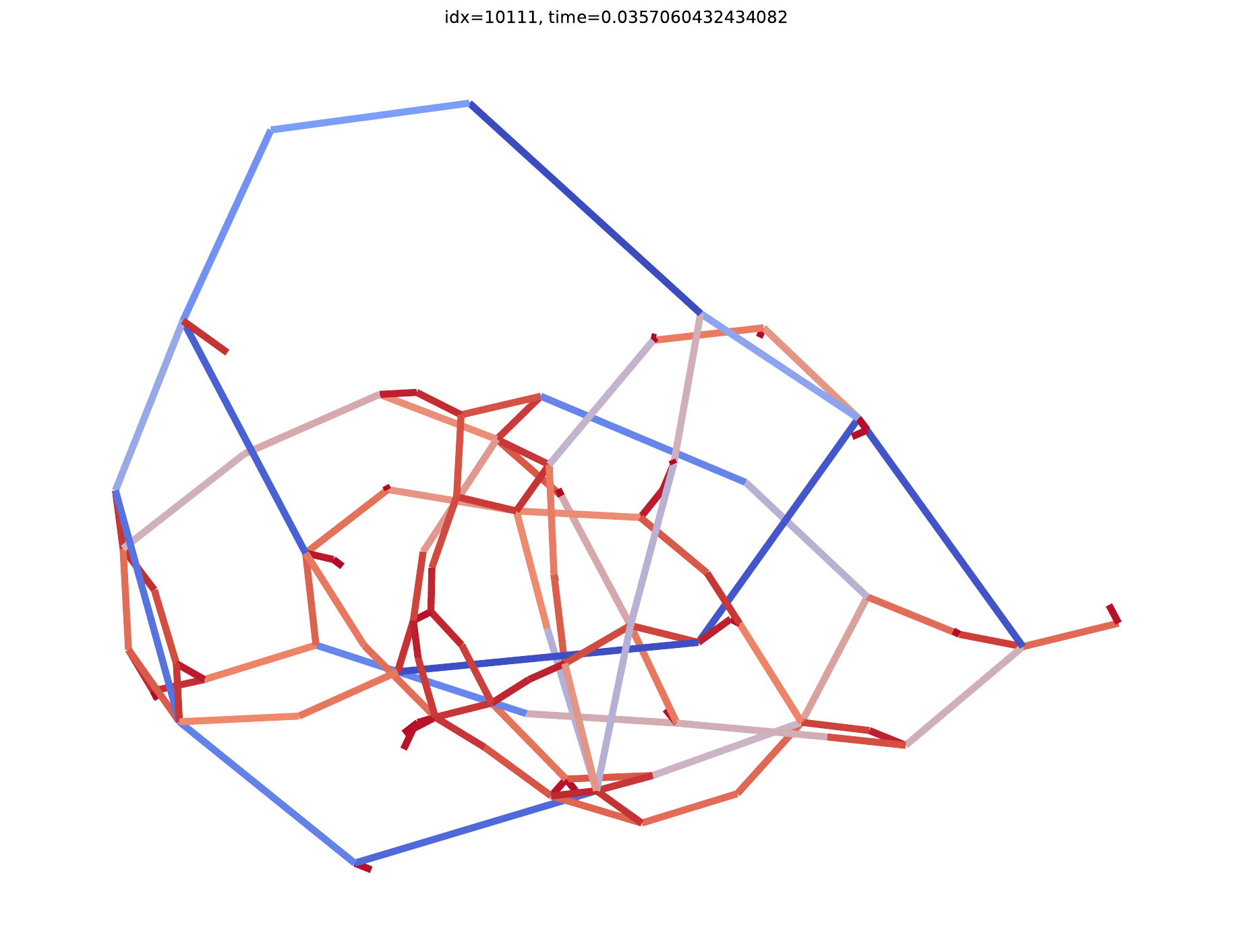} &
\imgcell{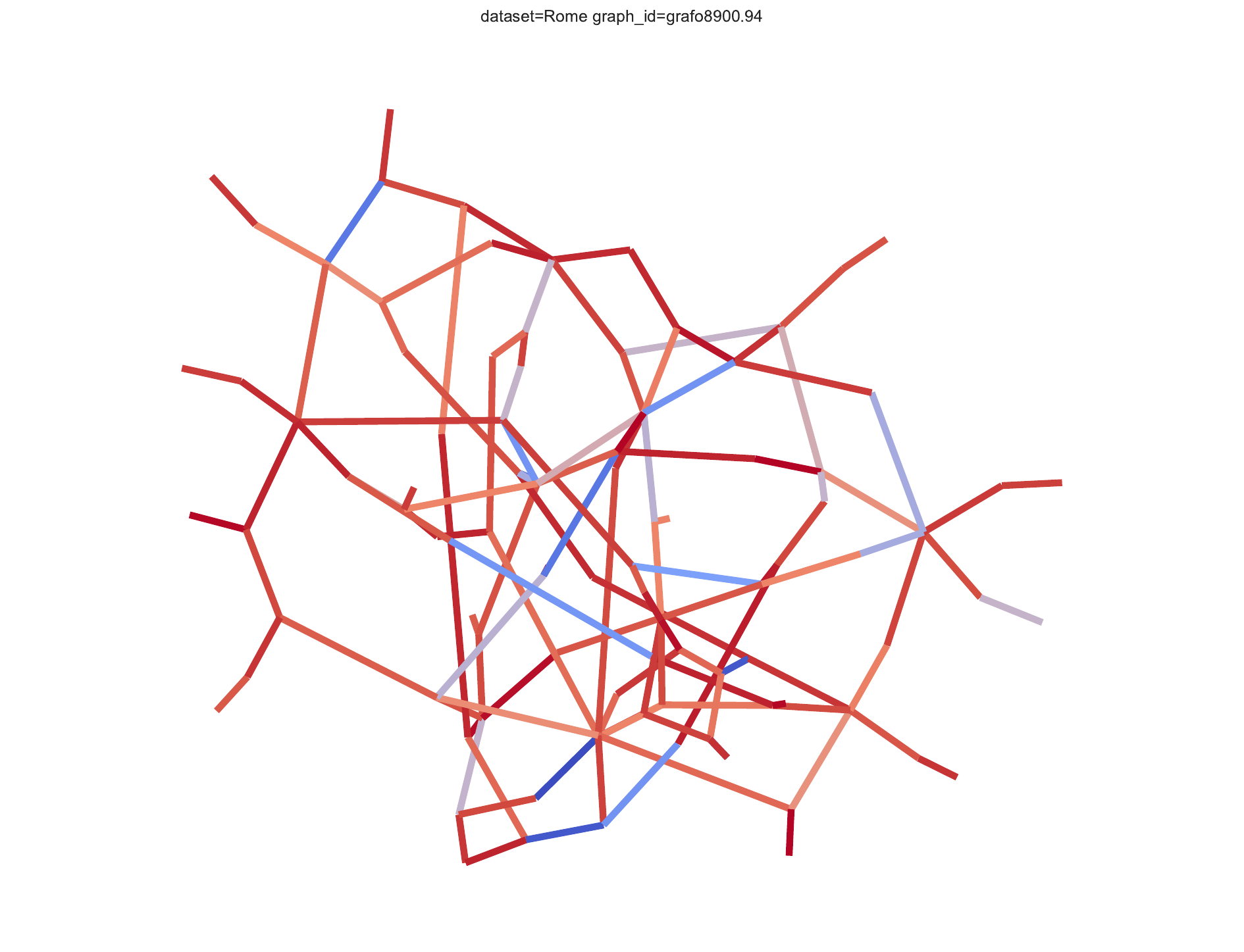} &
\imgcell{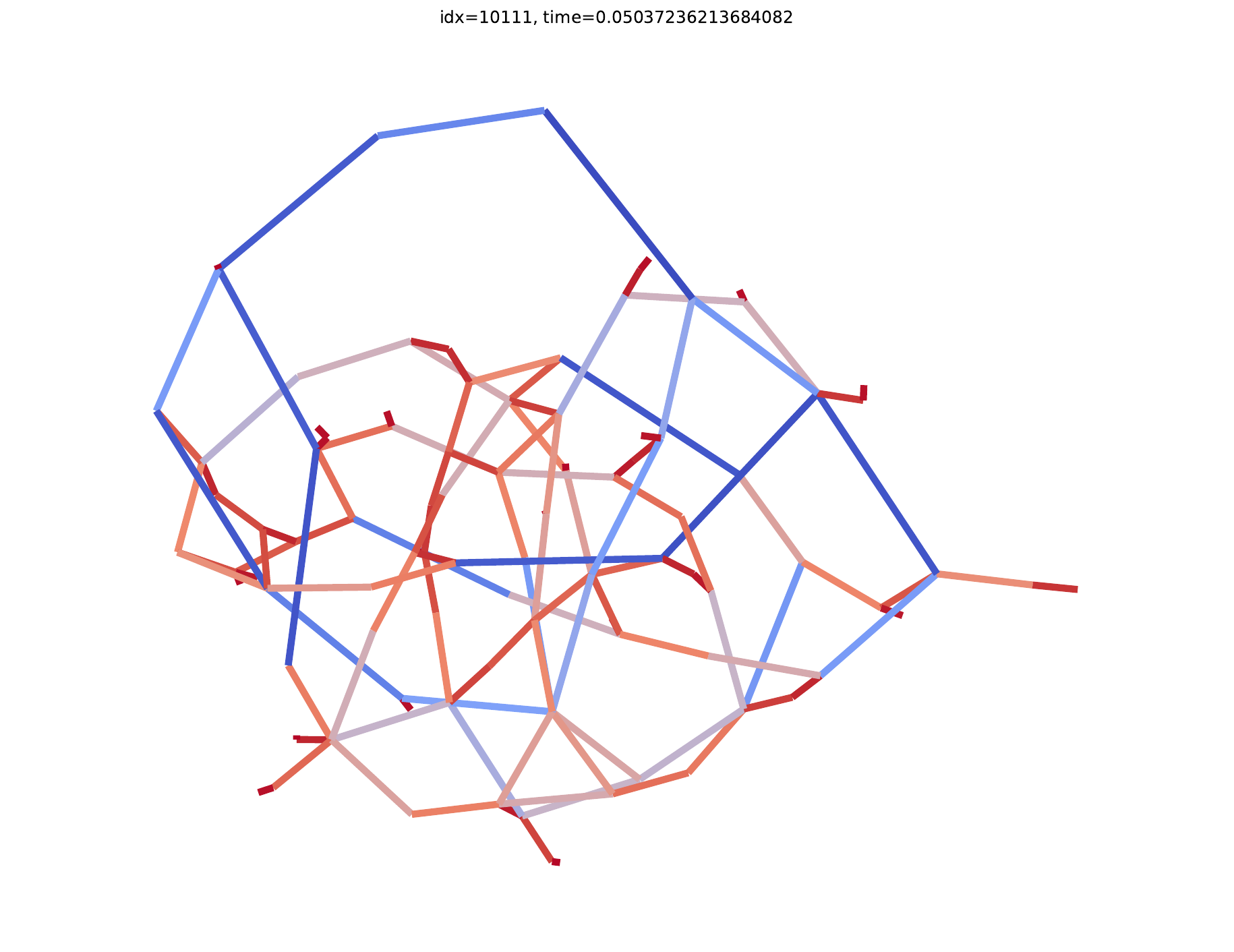} &
\imgcell{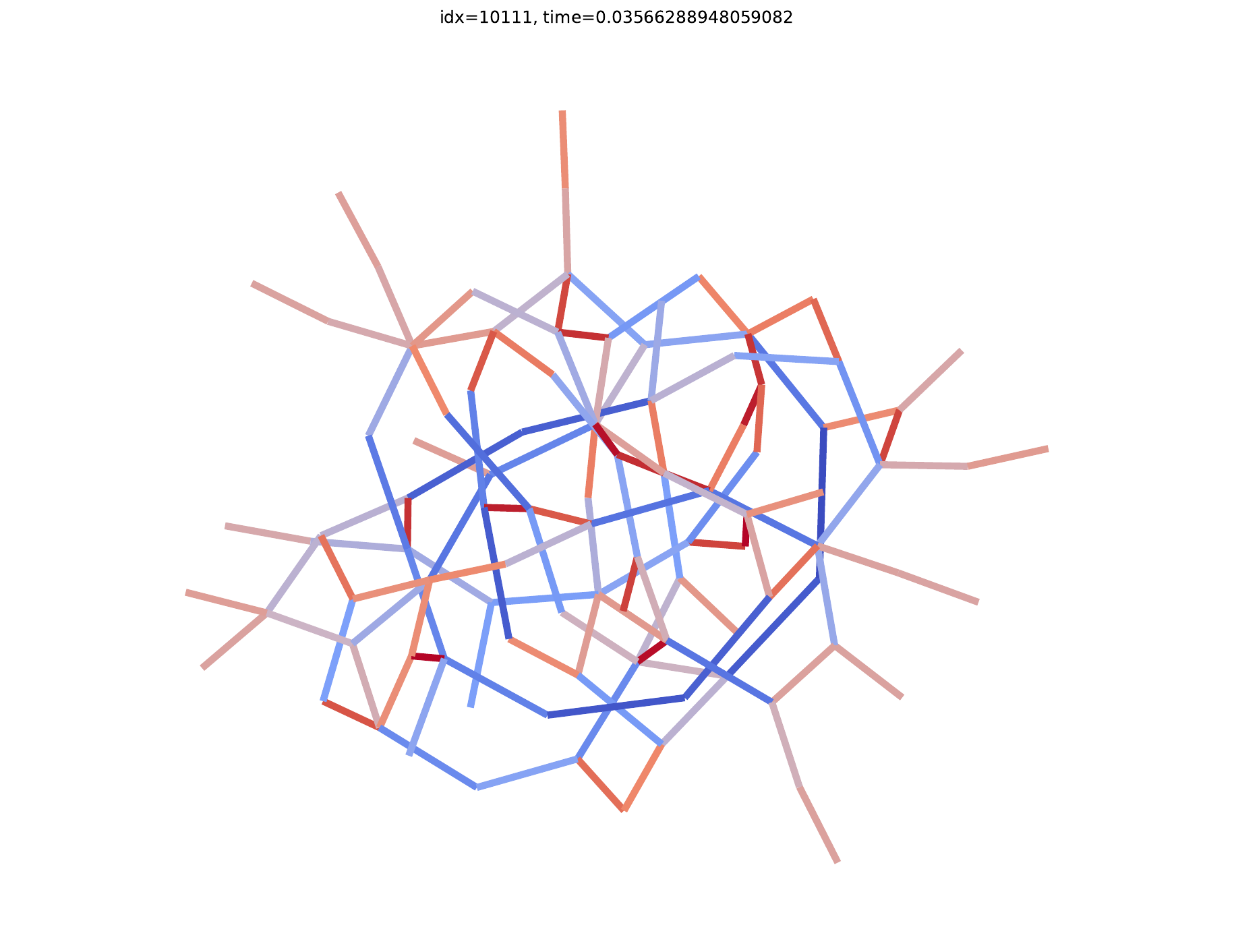} &
\imgcell{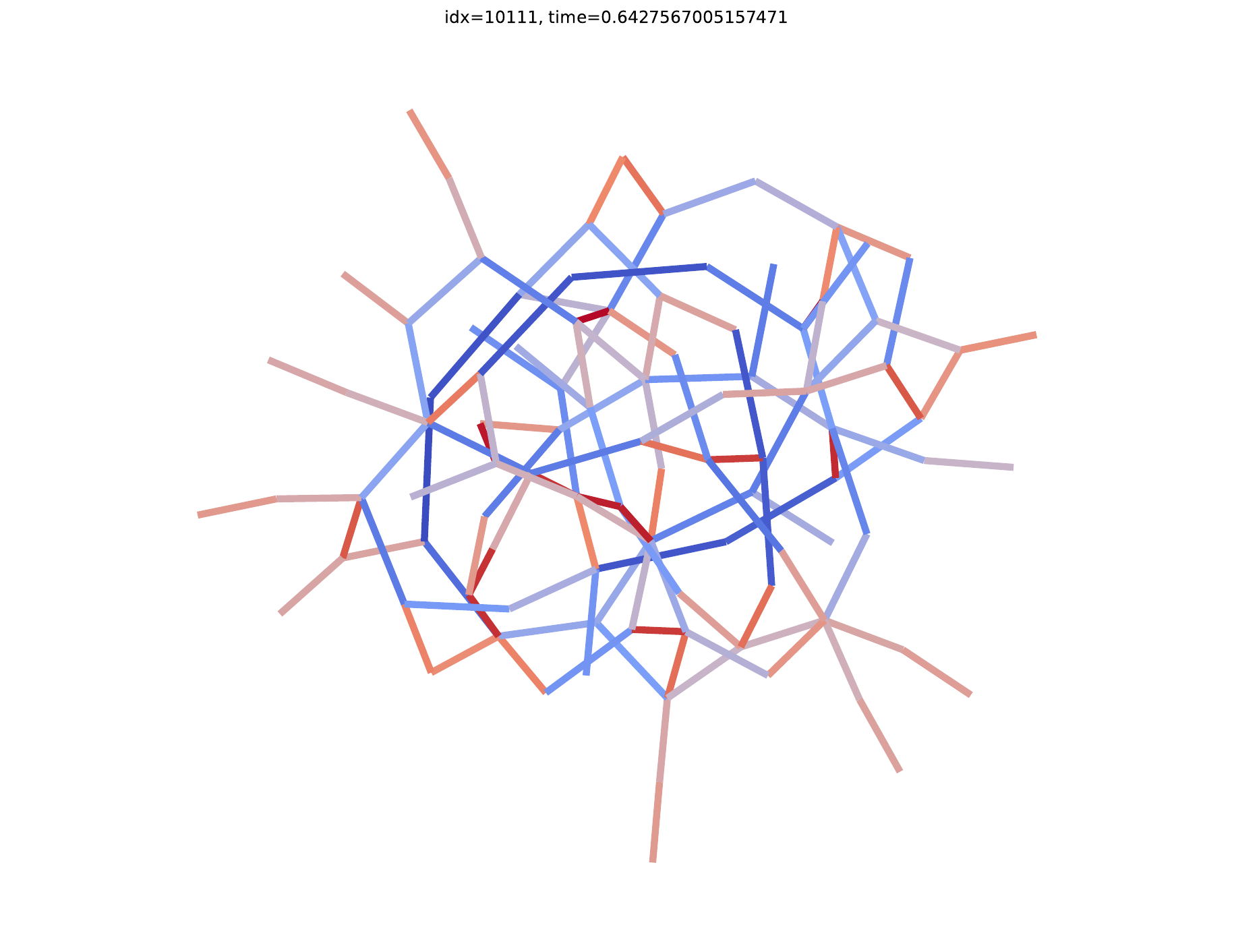} &
\imgcell{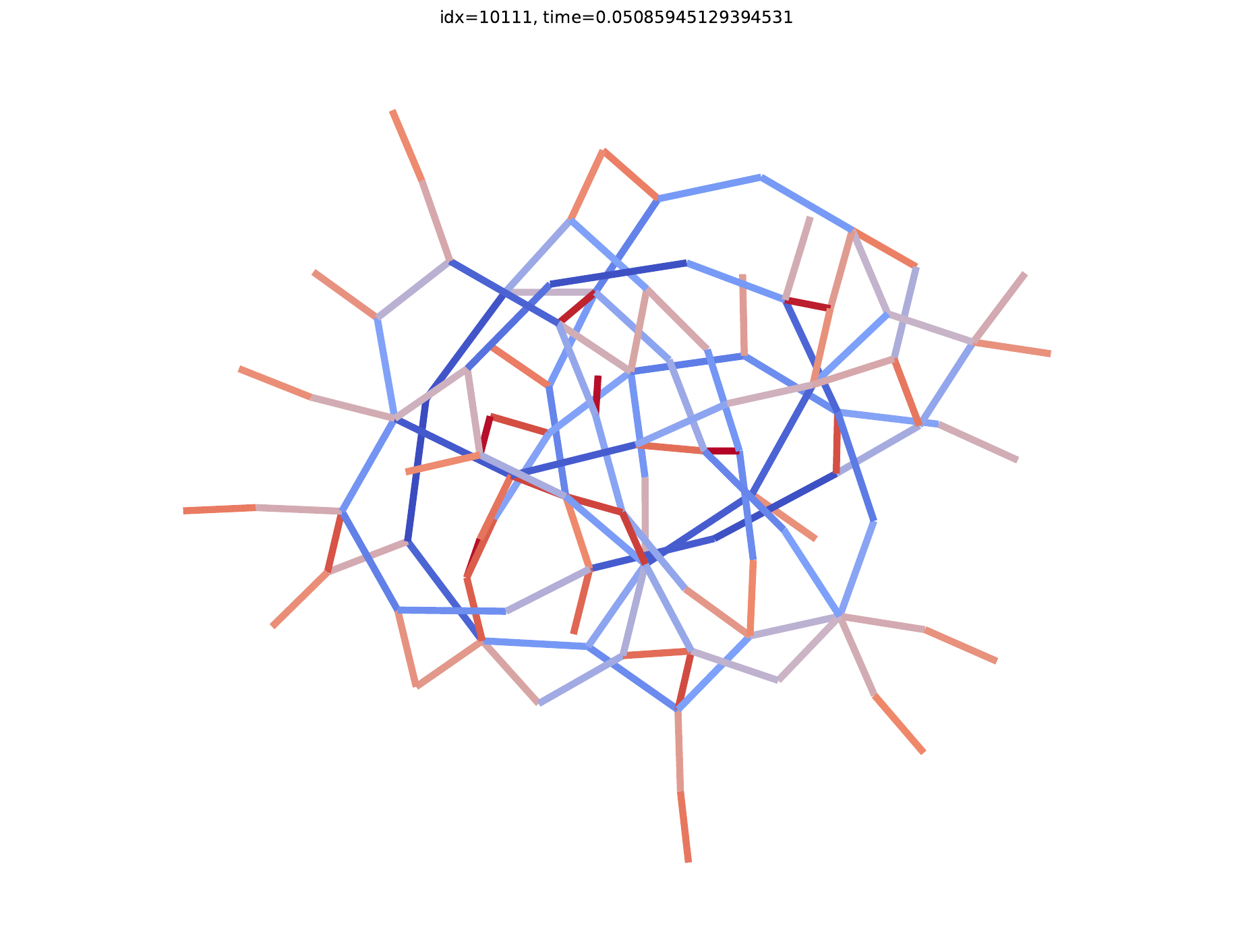} \\

&
t = 0.00s &
t = 0.78s &
t = 0.61s &
t = 0.05s &
t = 102.27s &
t = 0.04s &
t = 0.04s &
t = 0.05s &
t = 0.05s &
t = 0.04s &
t = 0.04s &
t = 0.05s \\

\makecell{\bfseries grafo9545.75\\N = 75\\M = 92} &
\imgcell{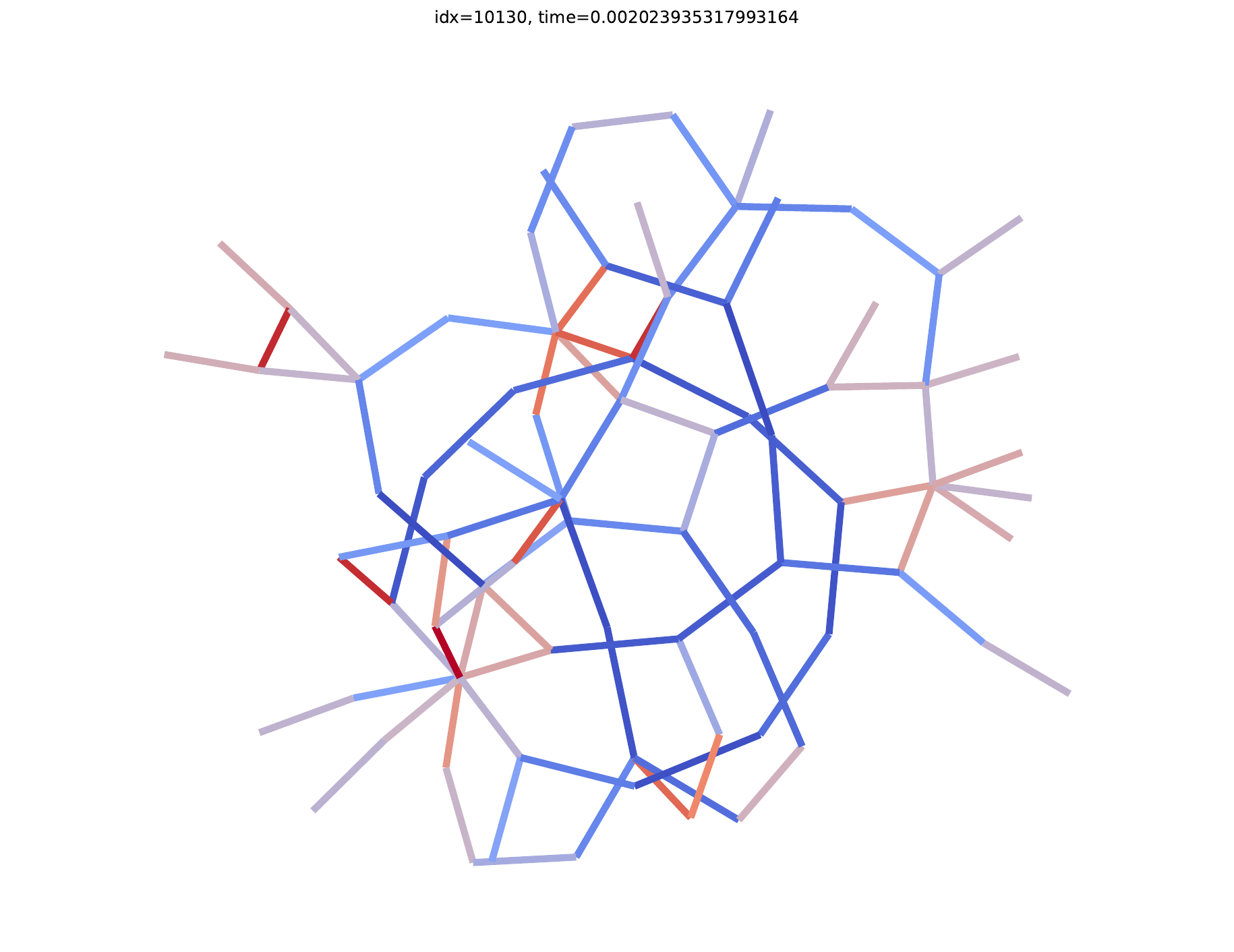} &
\imgcell{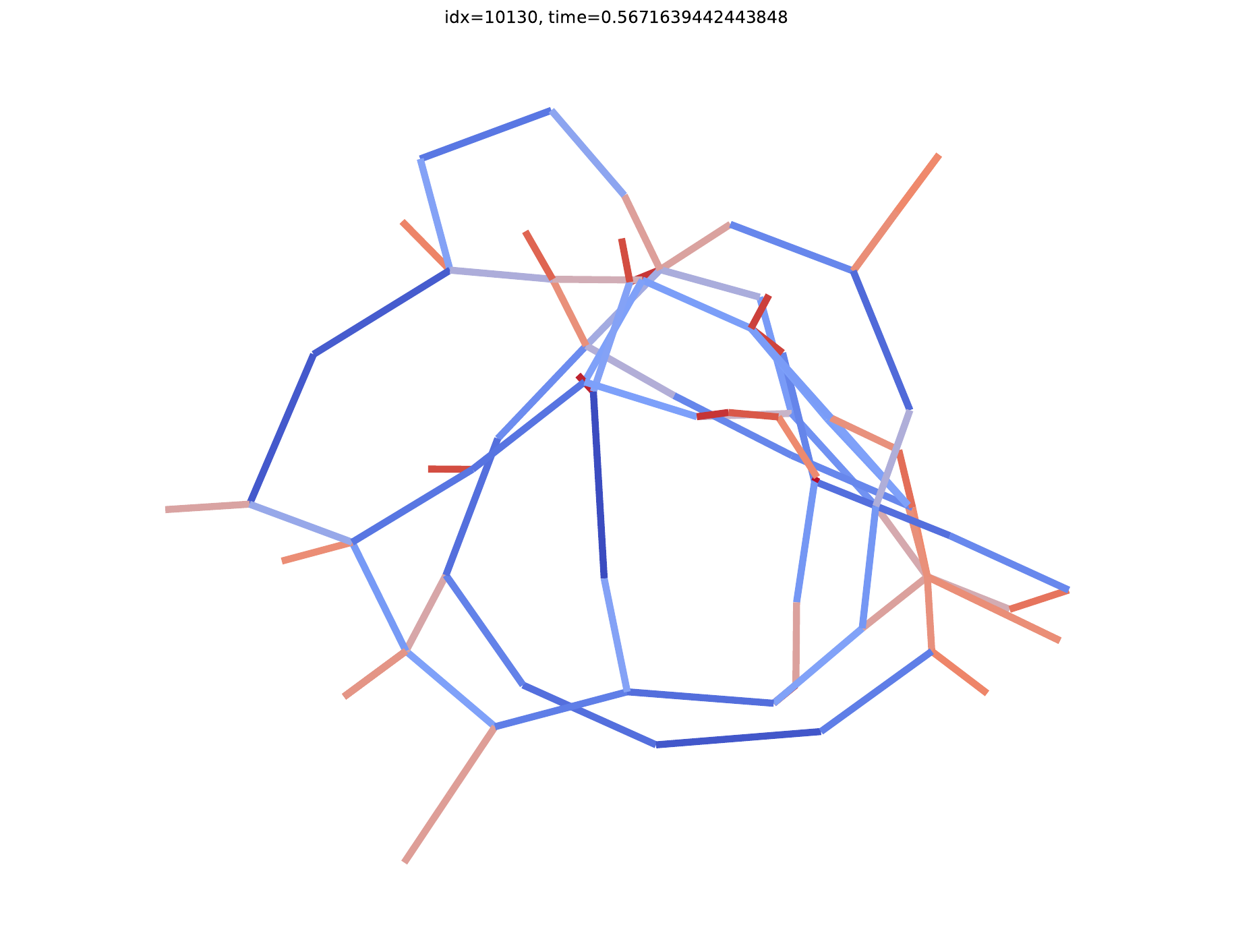} &
\imgcell{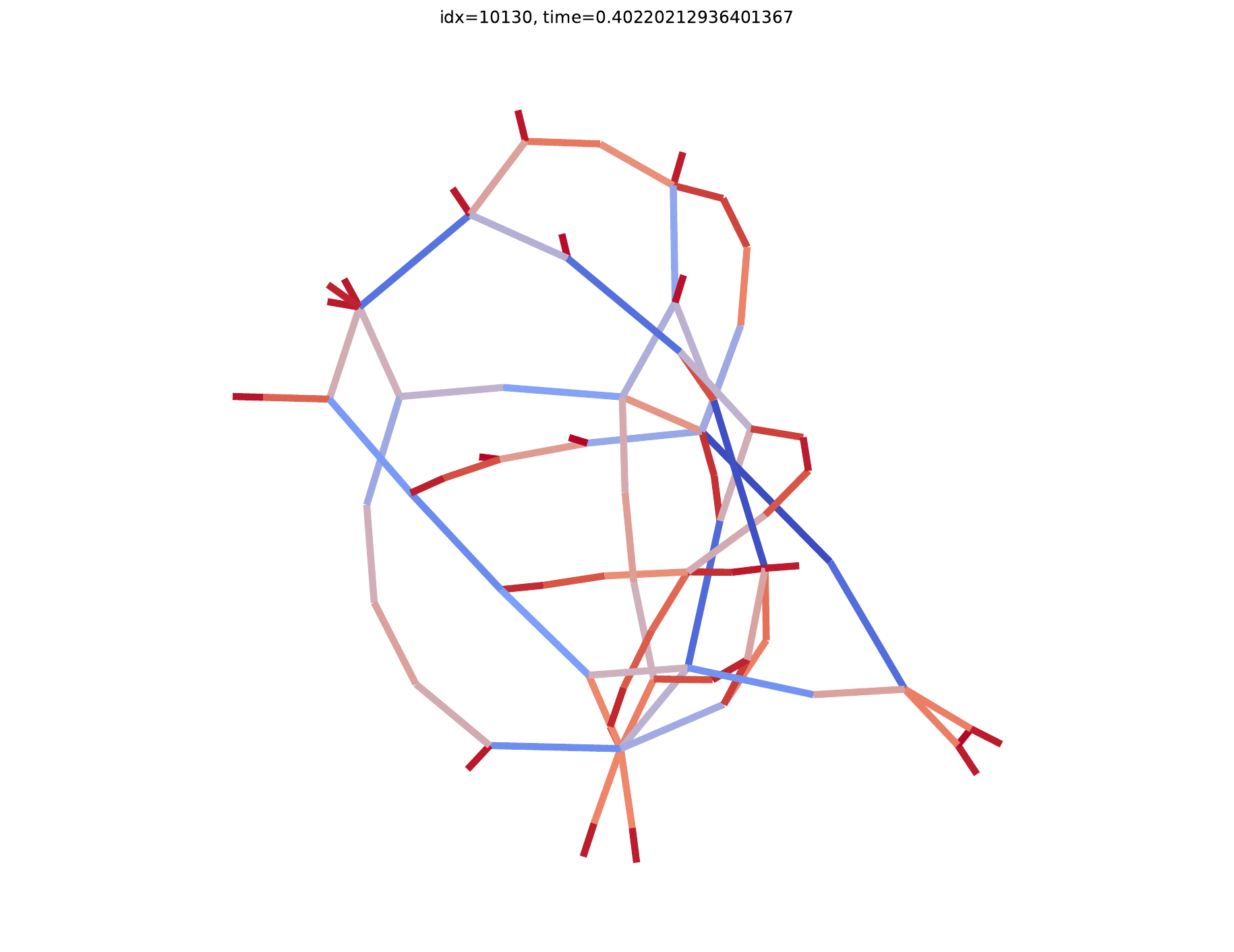} &
\imgcell{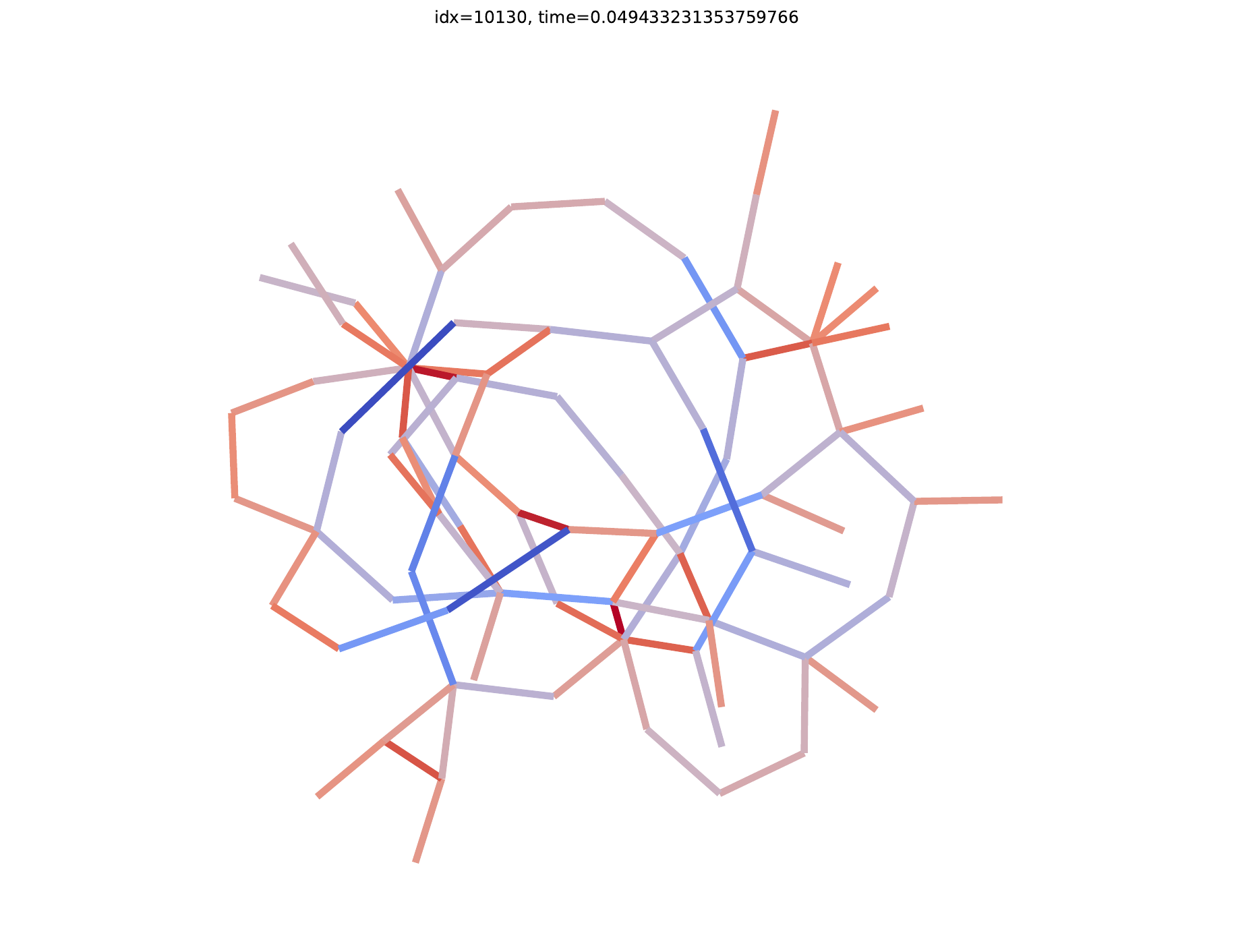} &
\imgcell{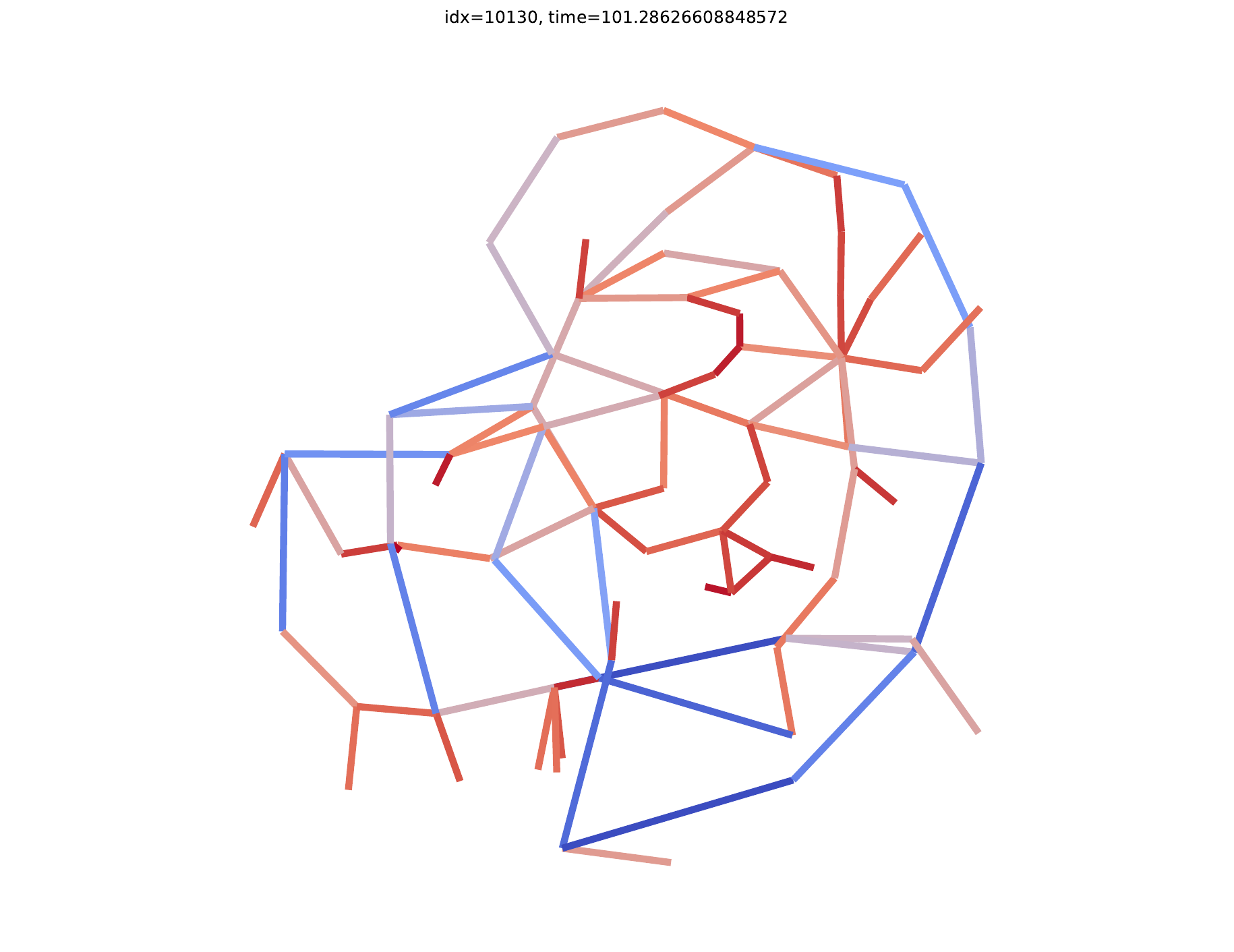} &
\imgcell{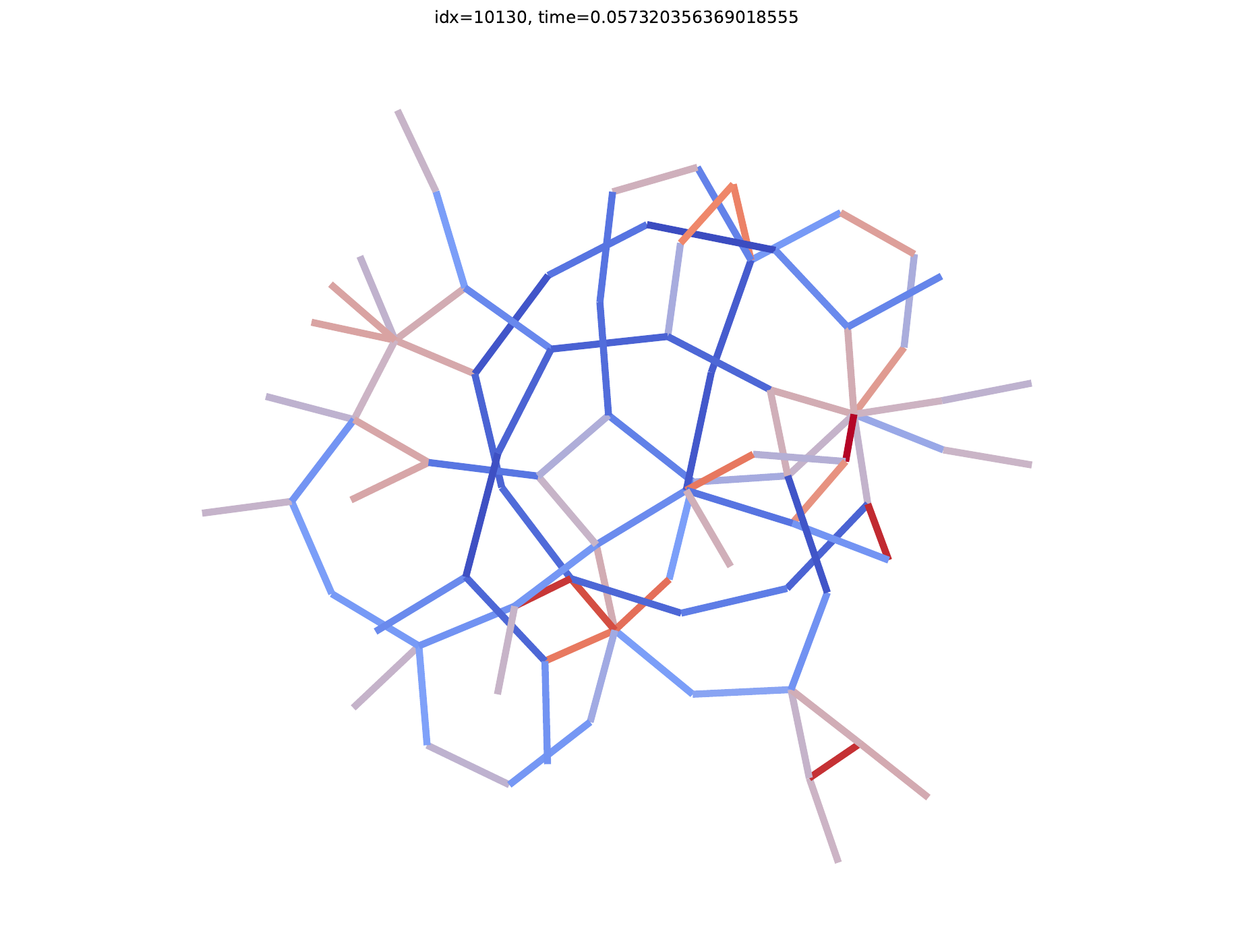} &
\imgcell{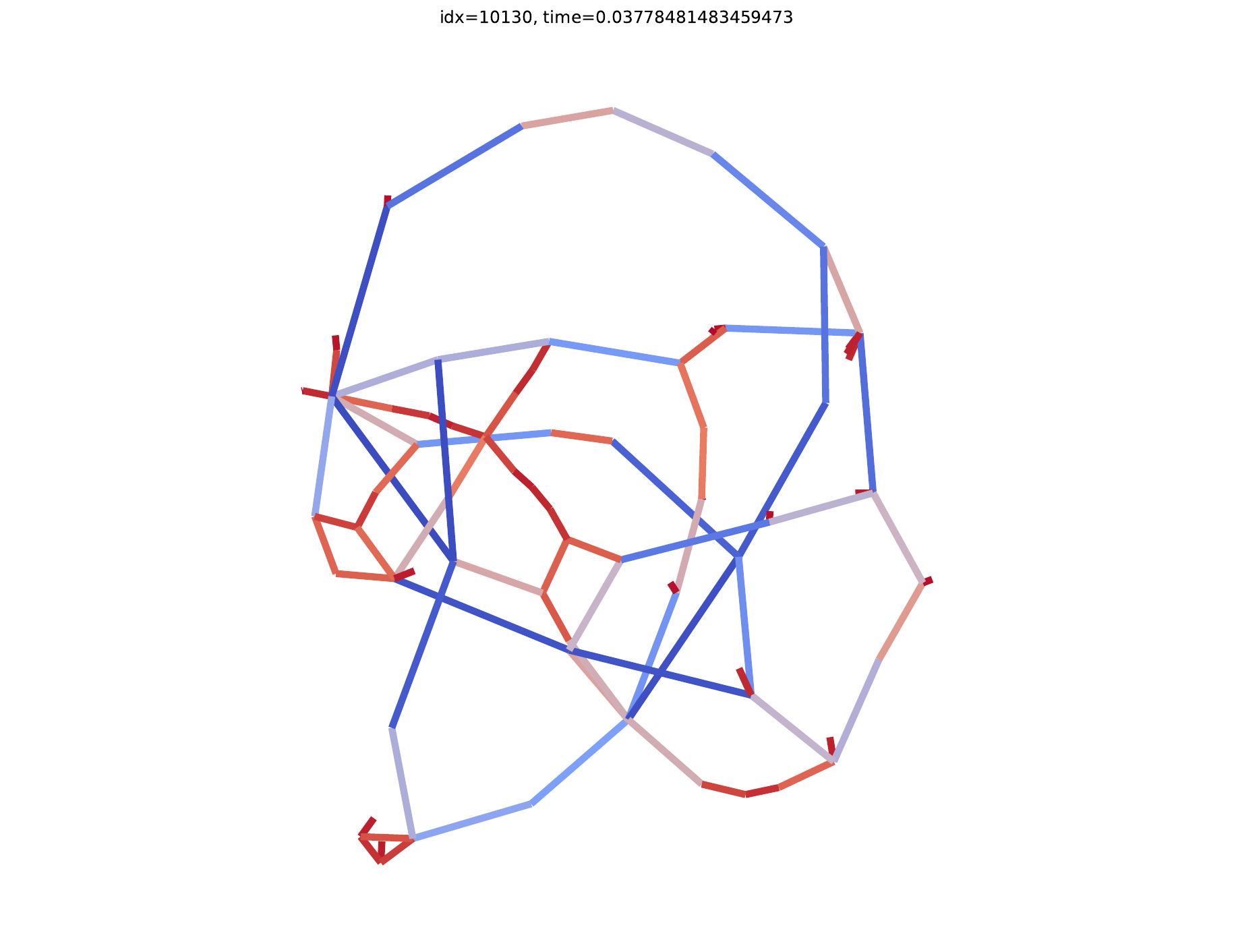} &
\imgcell{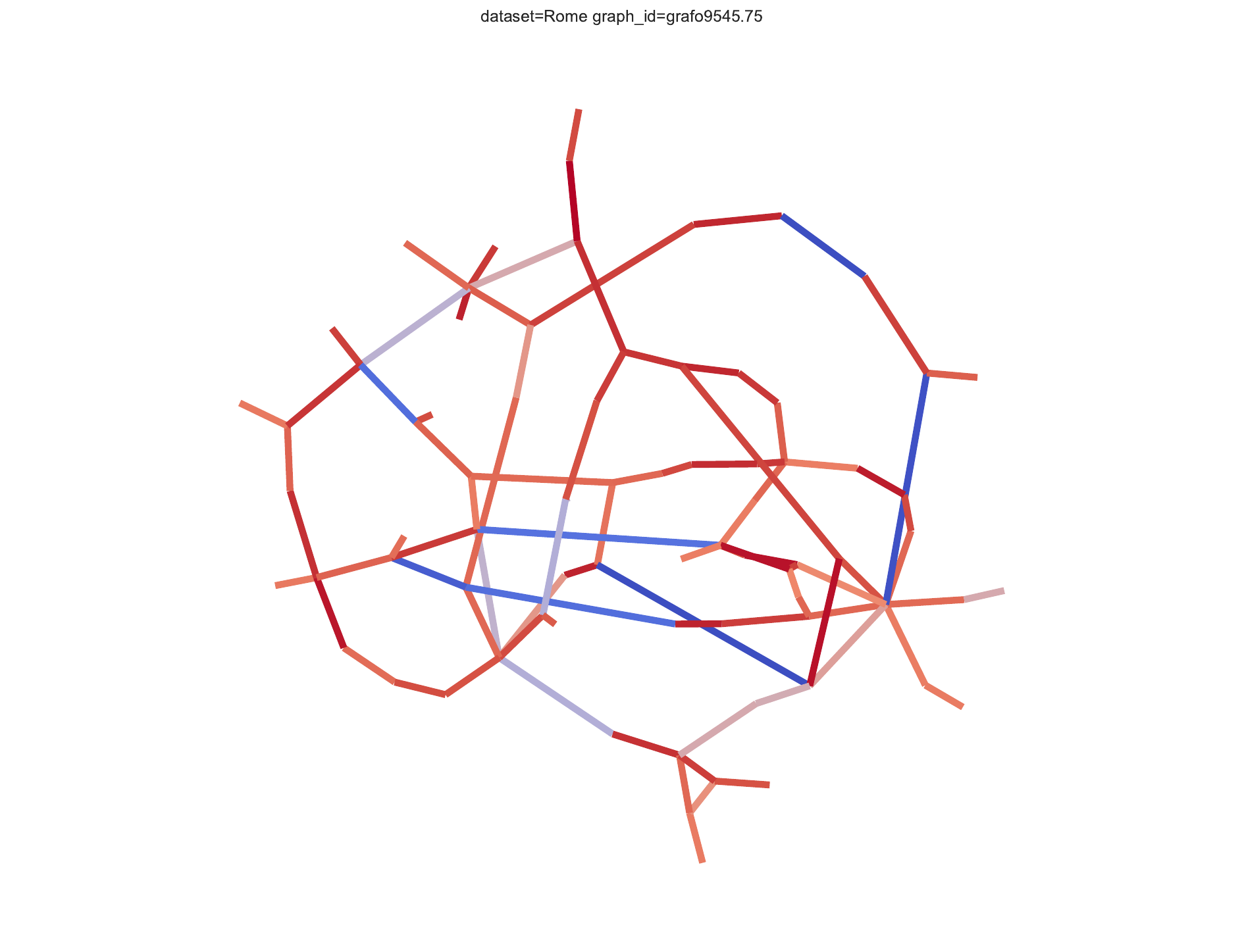} &
\imgcell{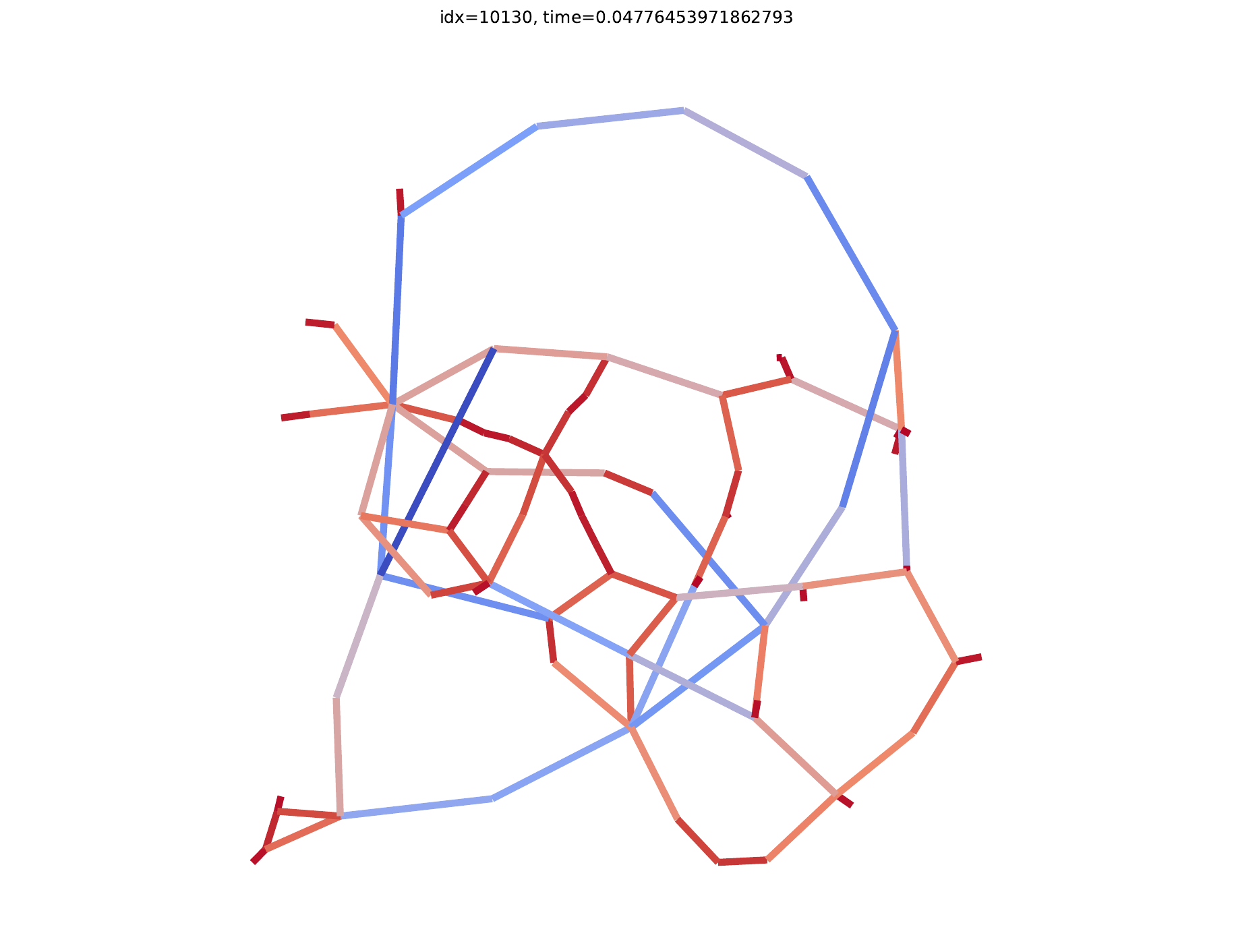} &
\imgcell{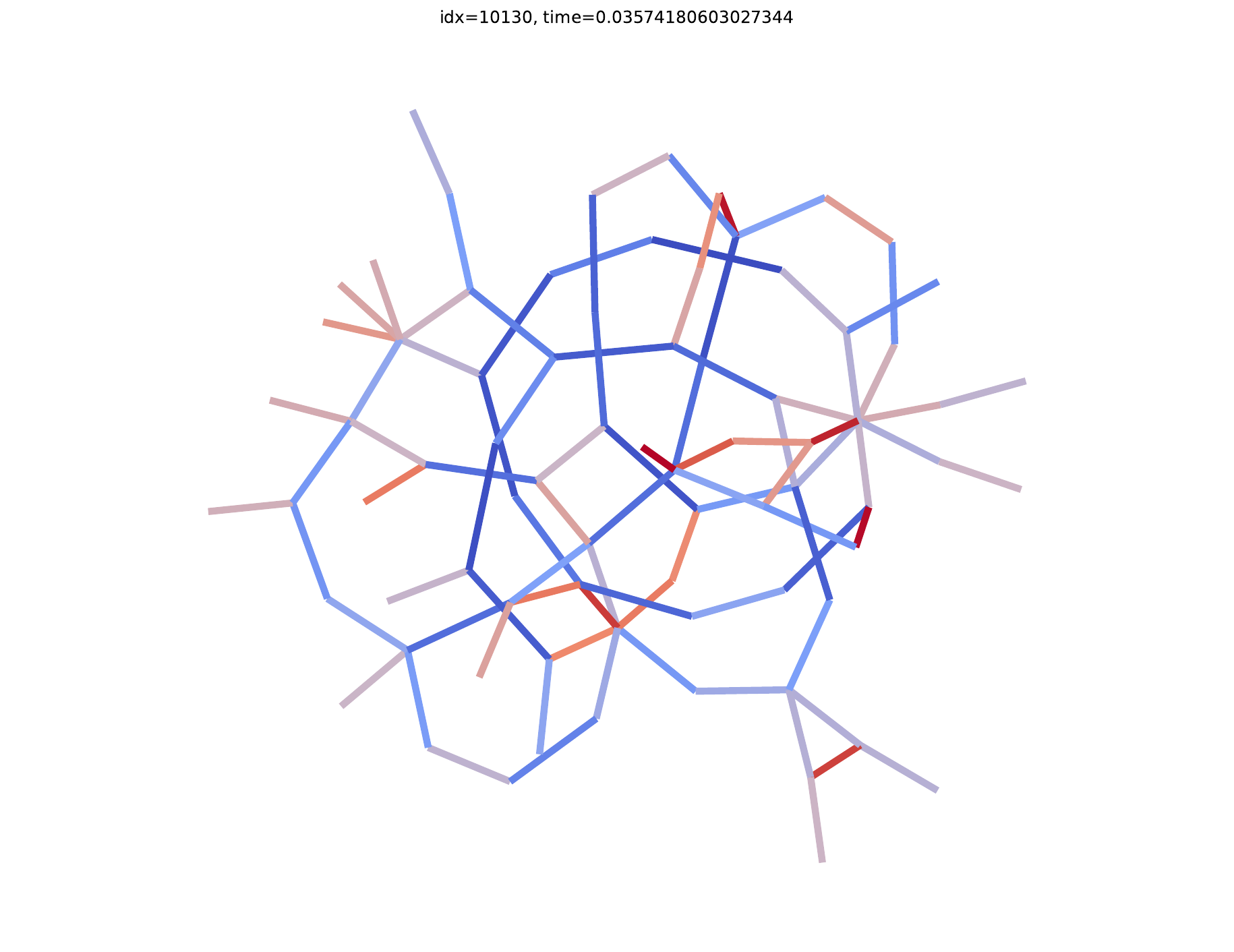} &
\imgcell{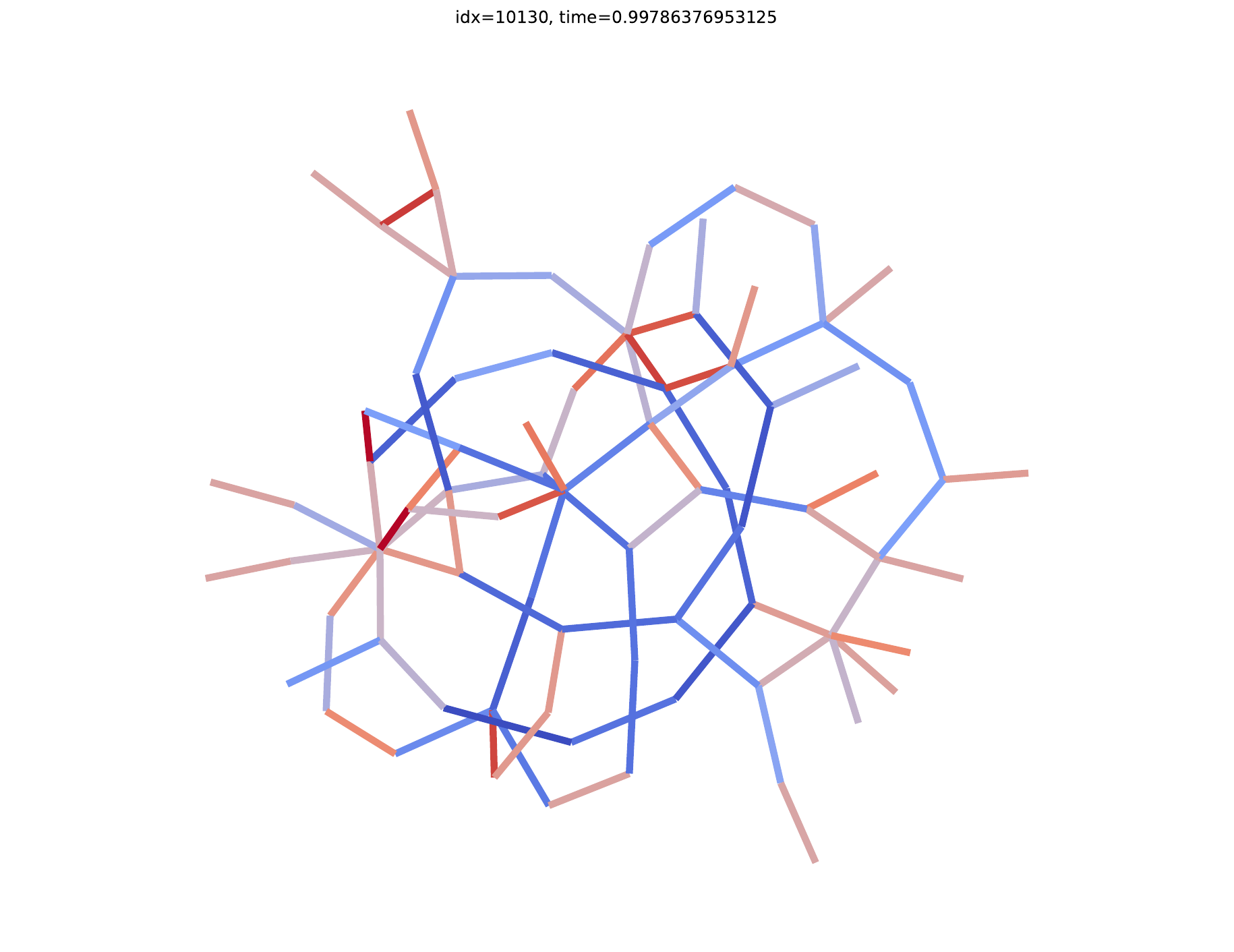} &
\imgcell{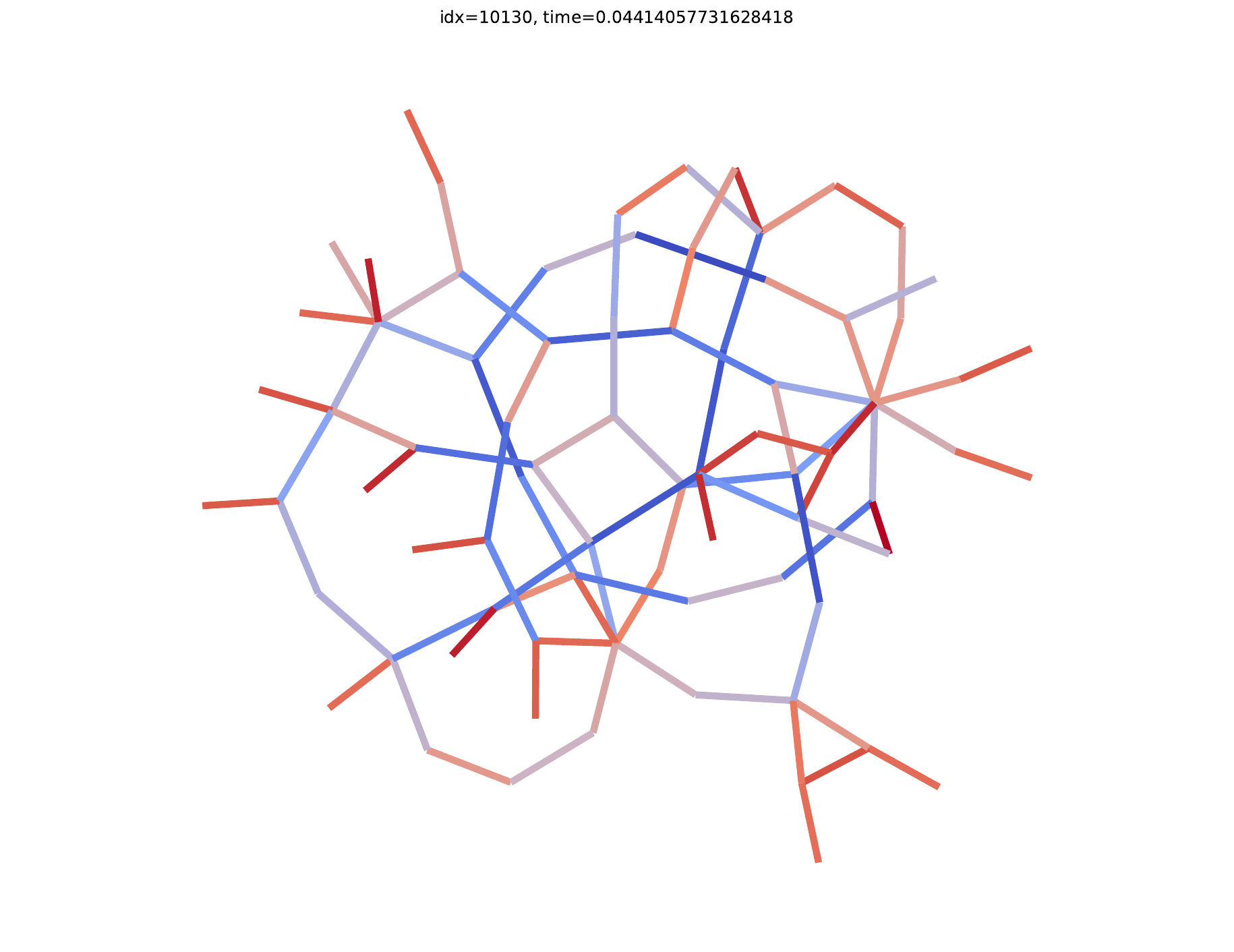} \\

&
t = 0.00s &
t = 0.57s &
t = 0.40s &
t = 0.05s &
t = 101.29s &
t = 0.06s &
t = 0.04s &
t = 0.05s &
t = 0.05s &
t = 0.04s &
t = 0.06s &
t = 0.04s \\

\makecell{\bfseries grafo5807.32\\N = 32\\M = 41} &
\imgcell{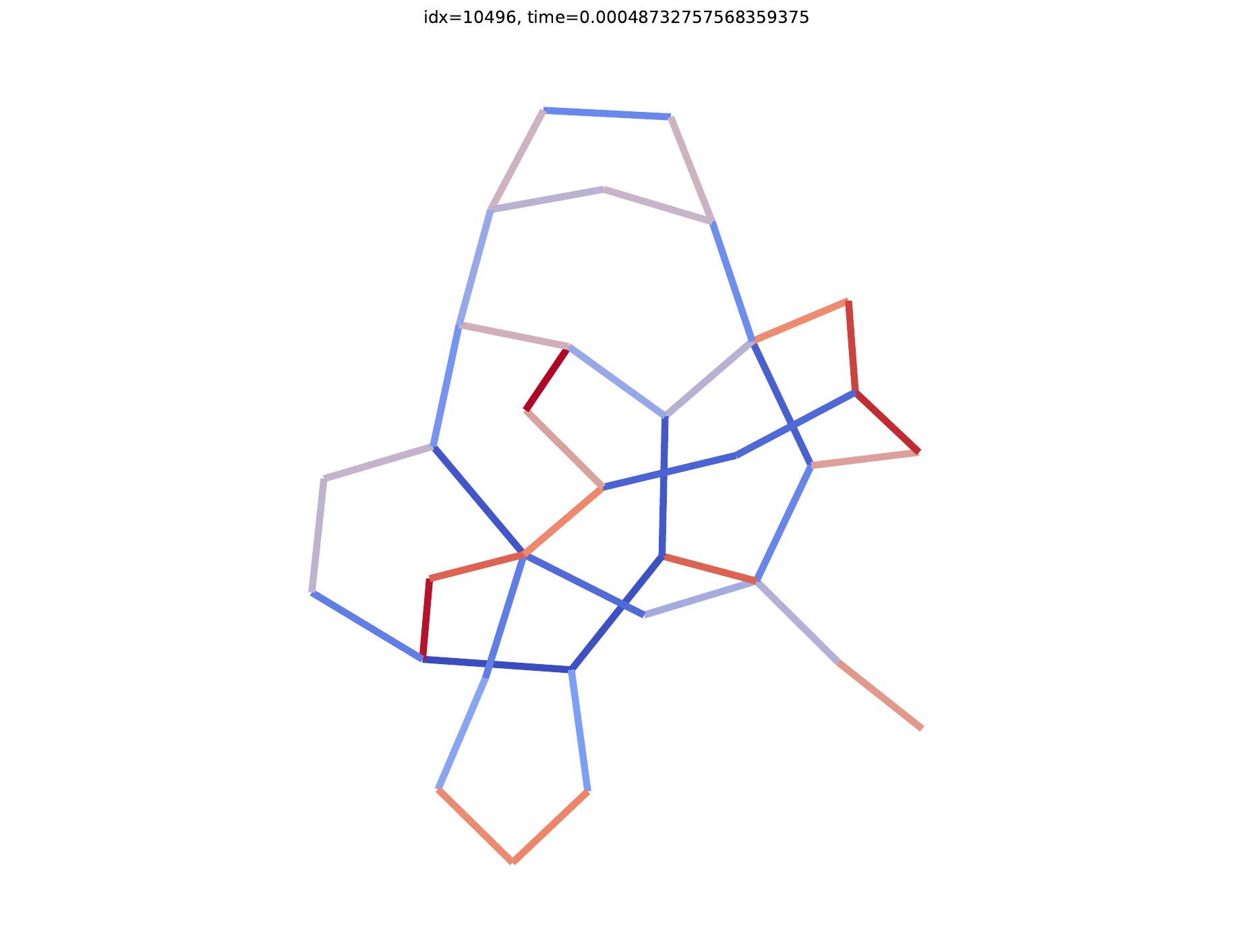} &
\imgcell{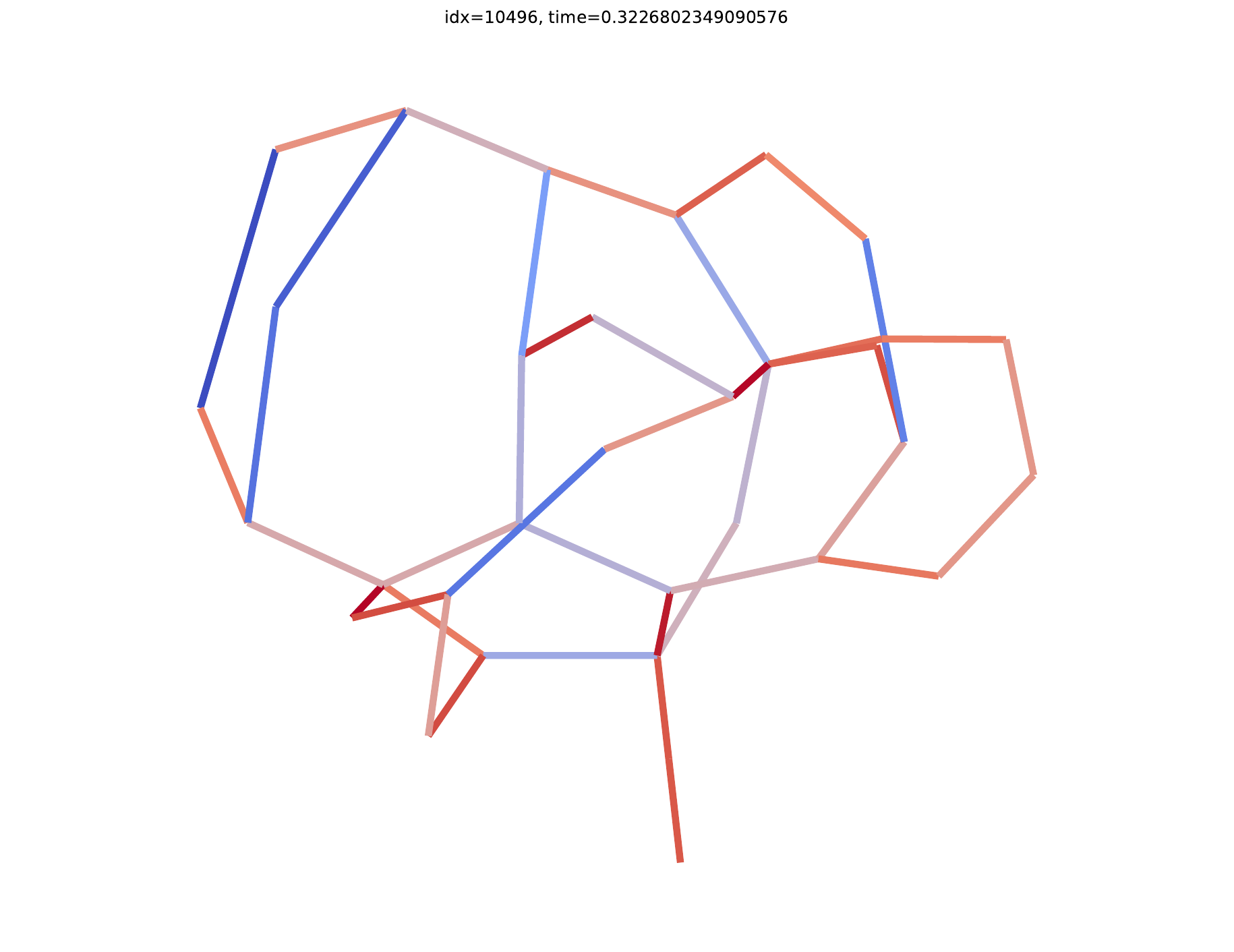} &
\imgcell{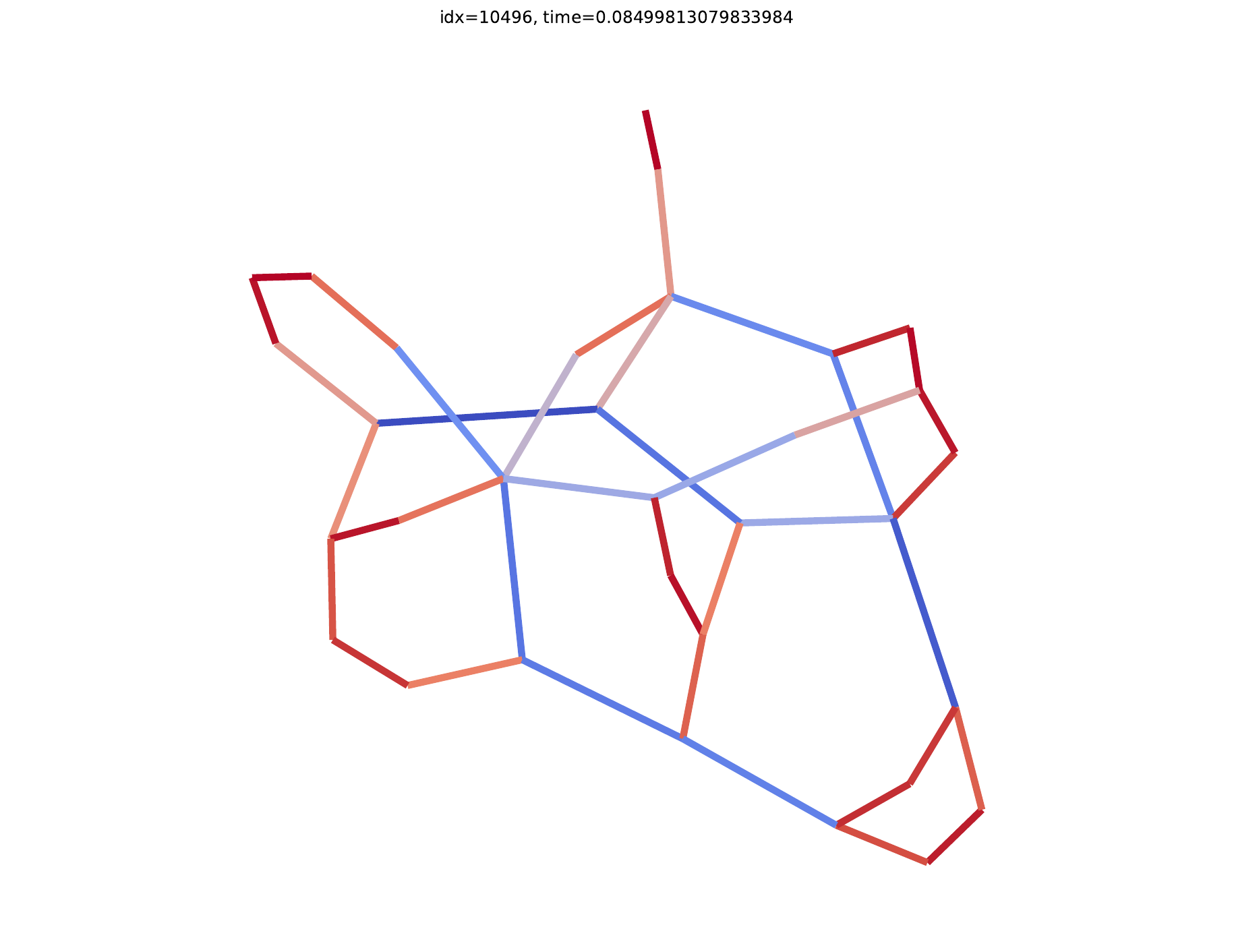} &
\imgcell{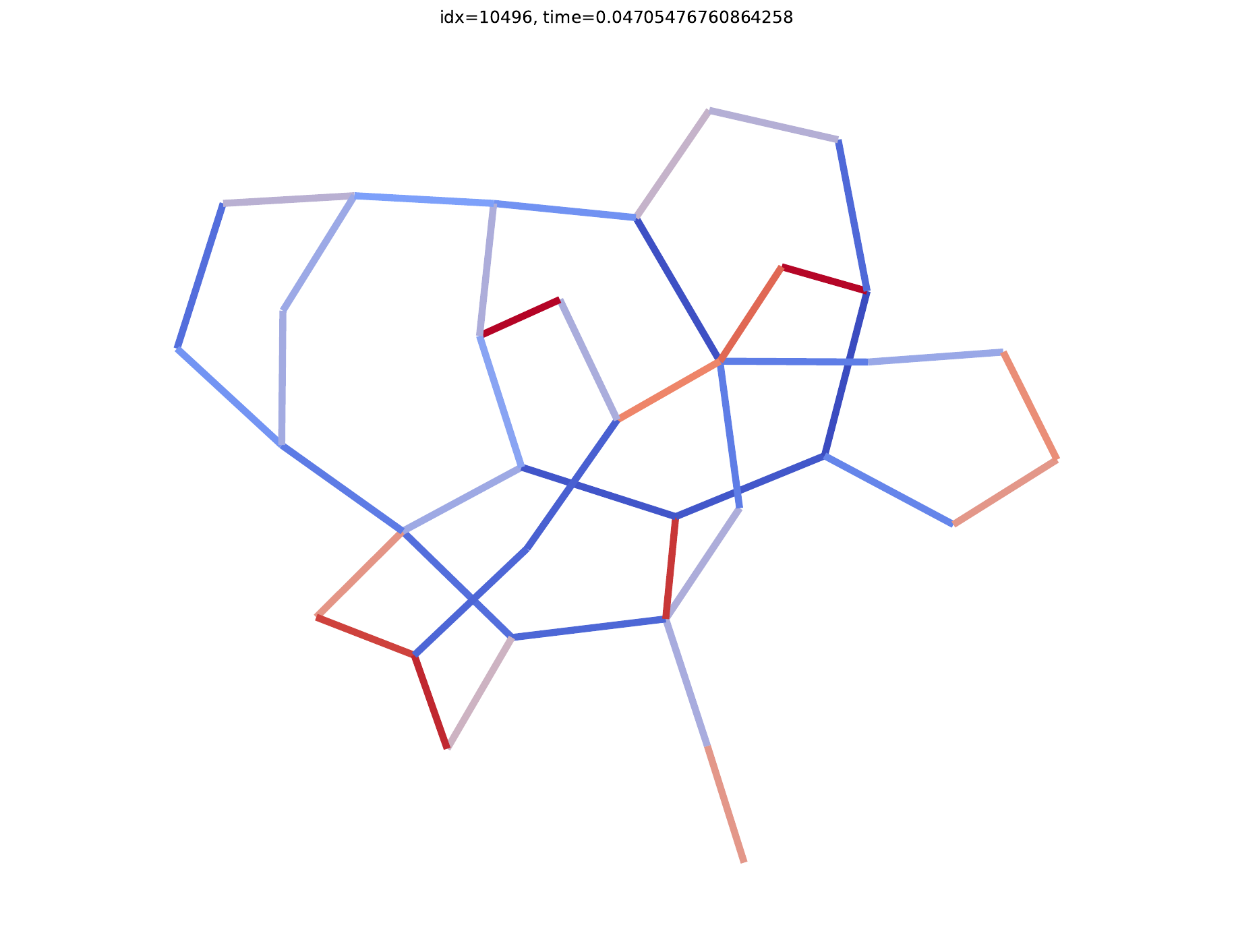} &
\imgcell{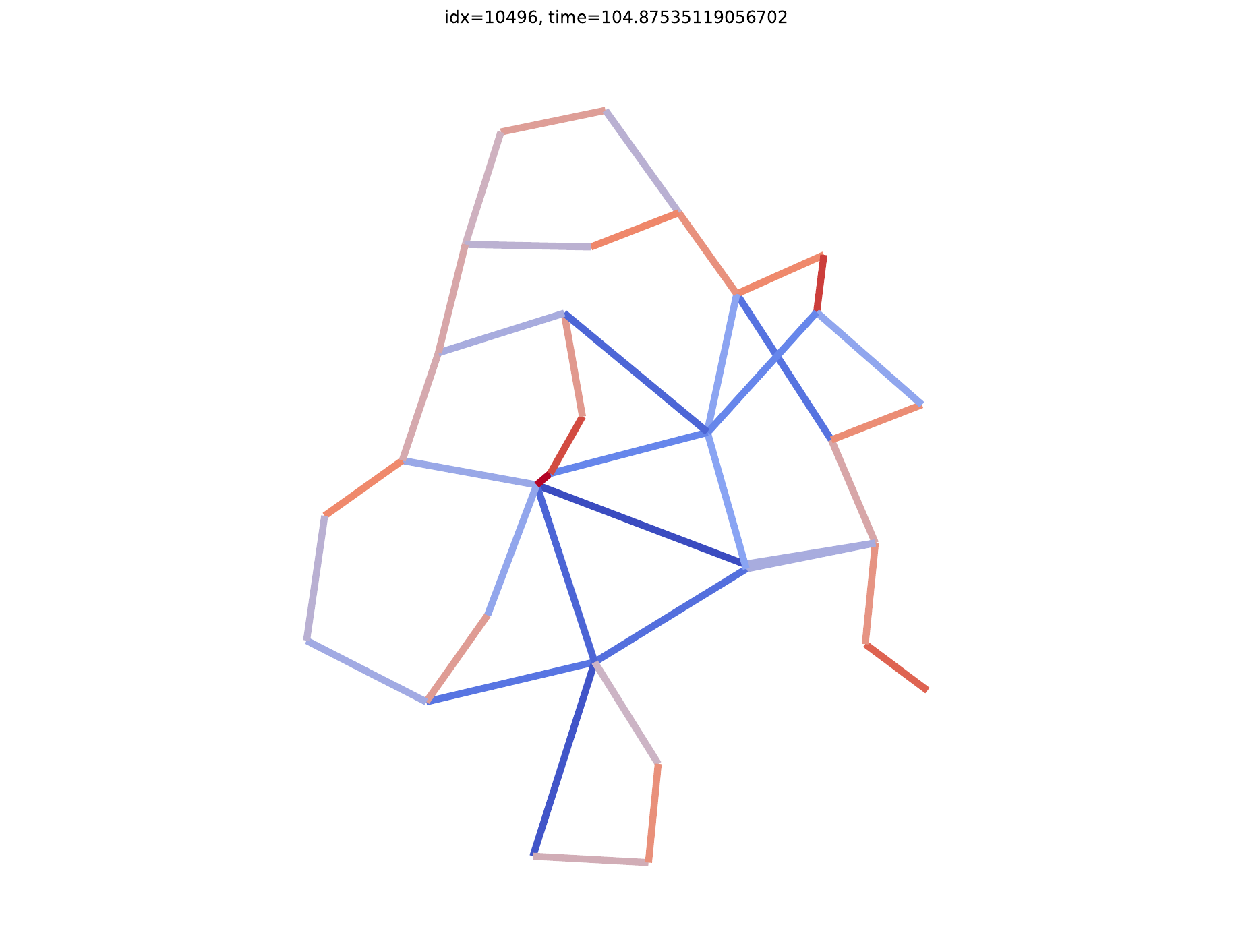} &
\imgcell{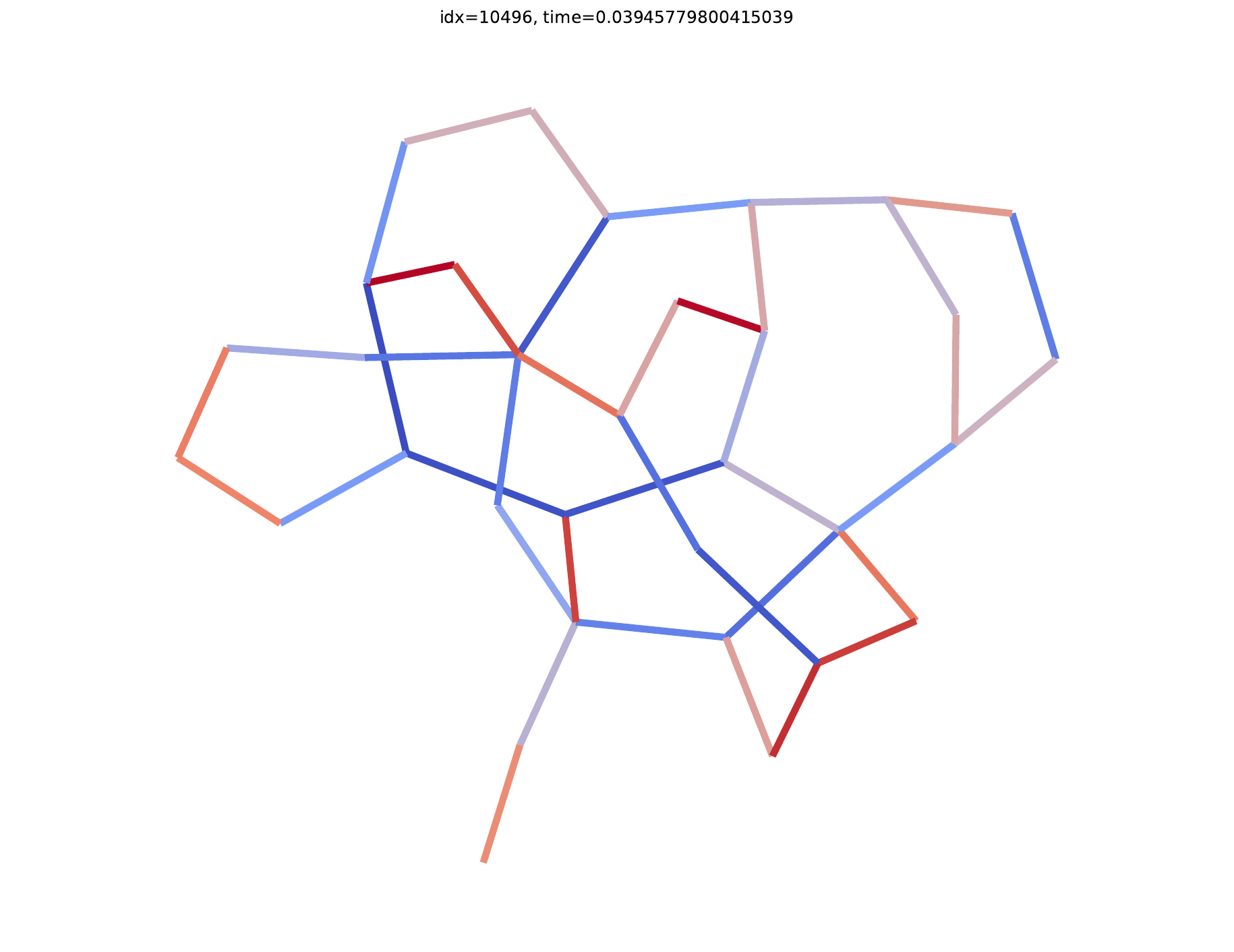} &
\imgcell{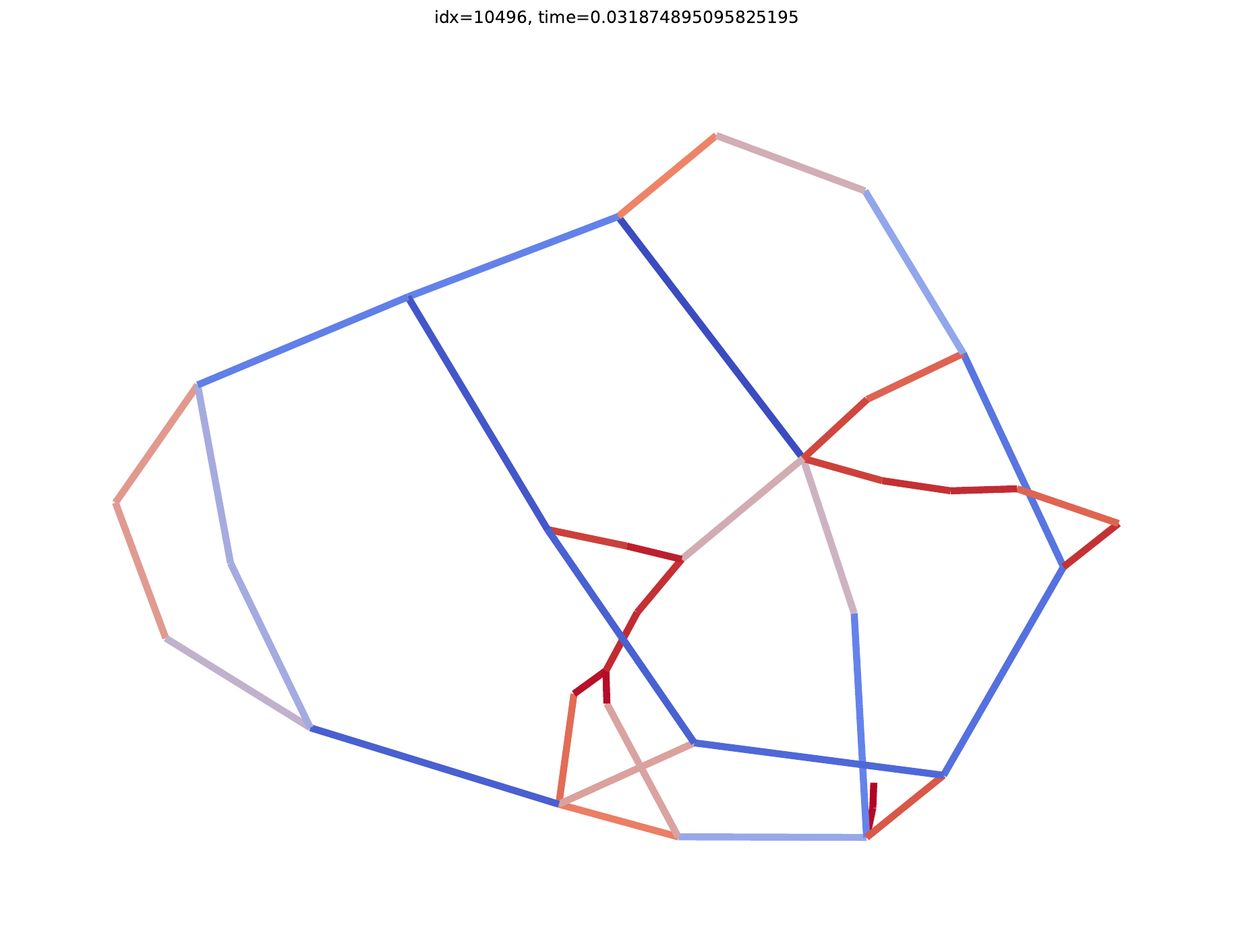} &
\imgcell{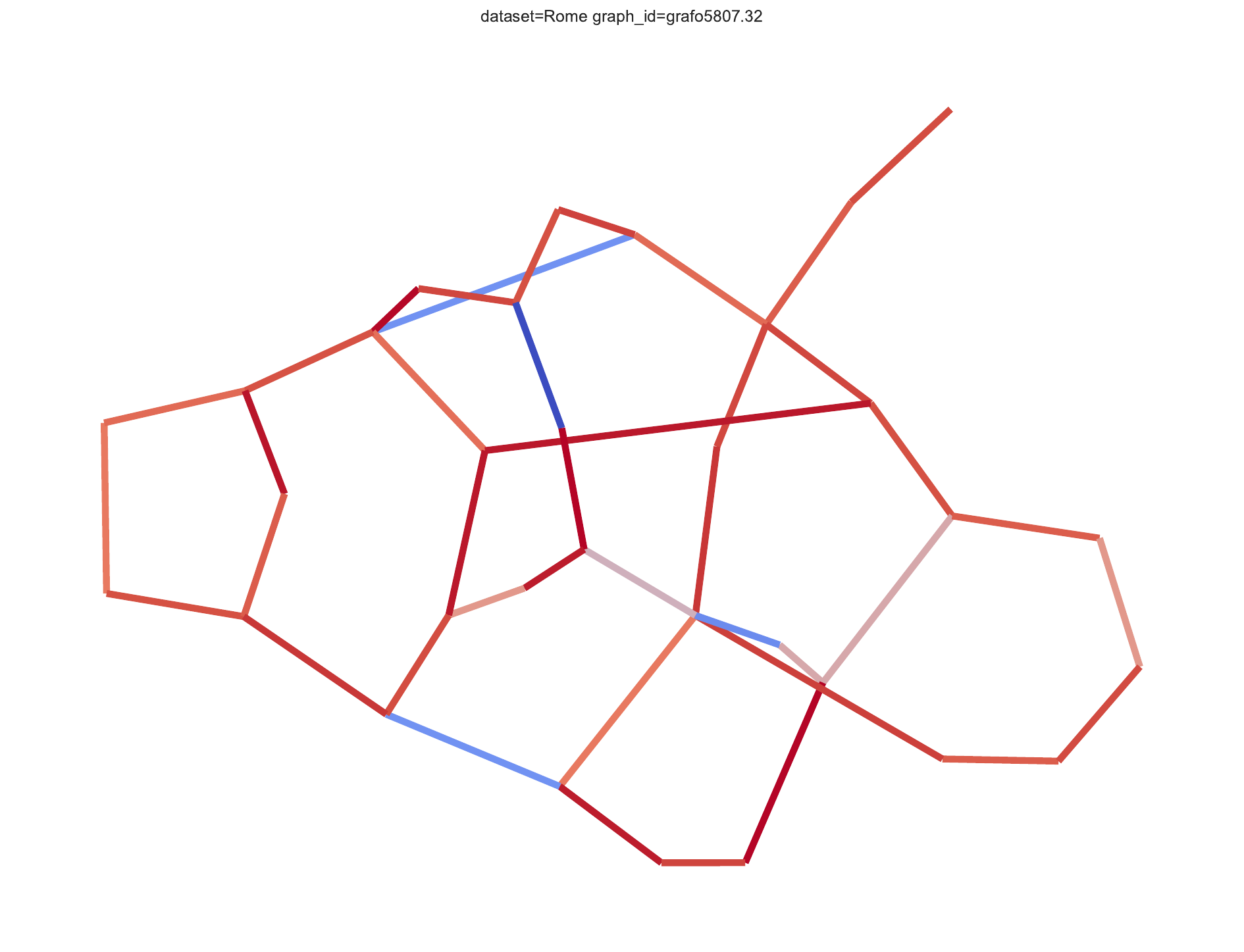} &
\imgcell{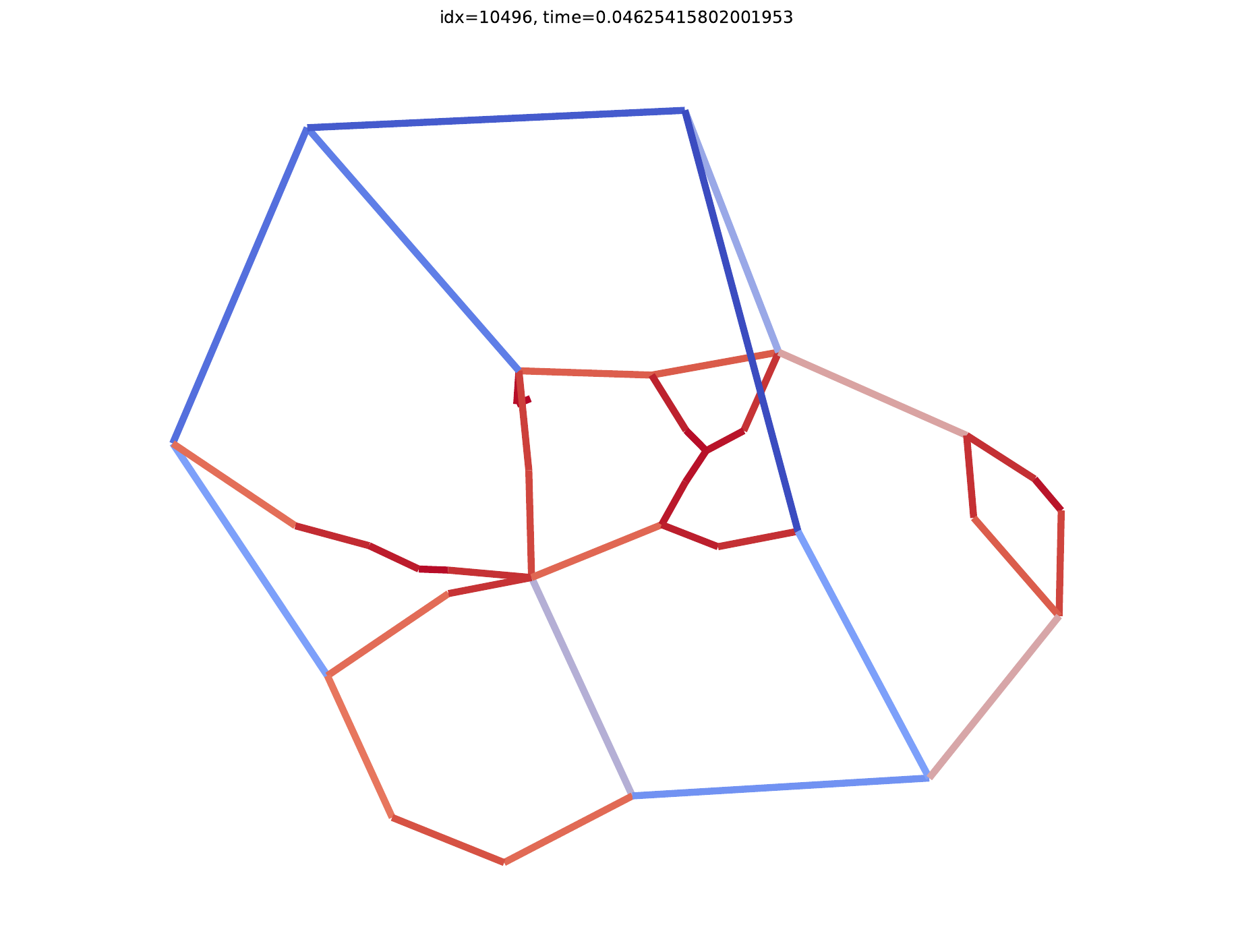} &
\imgcell{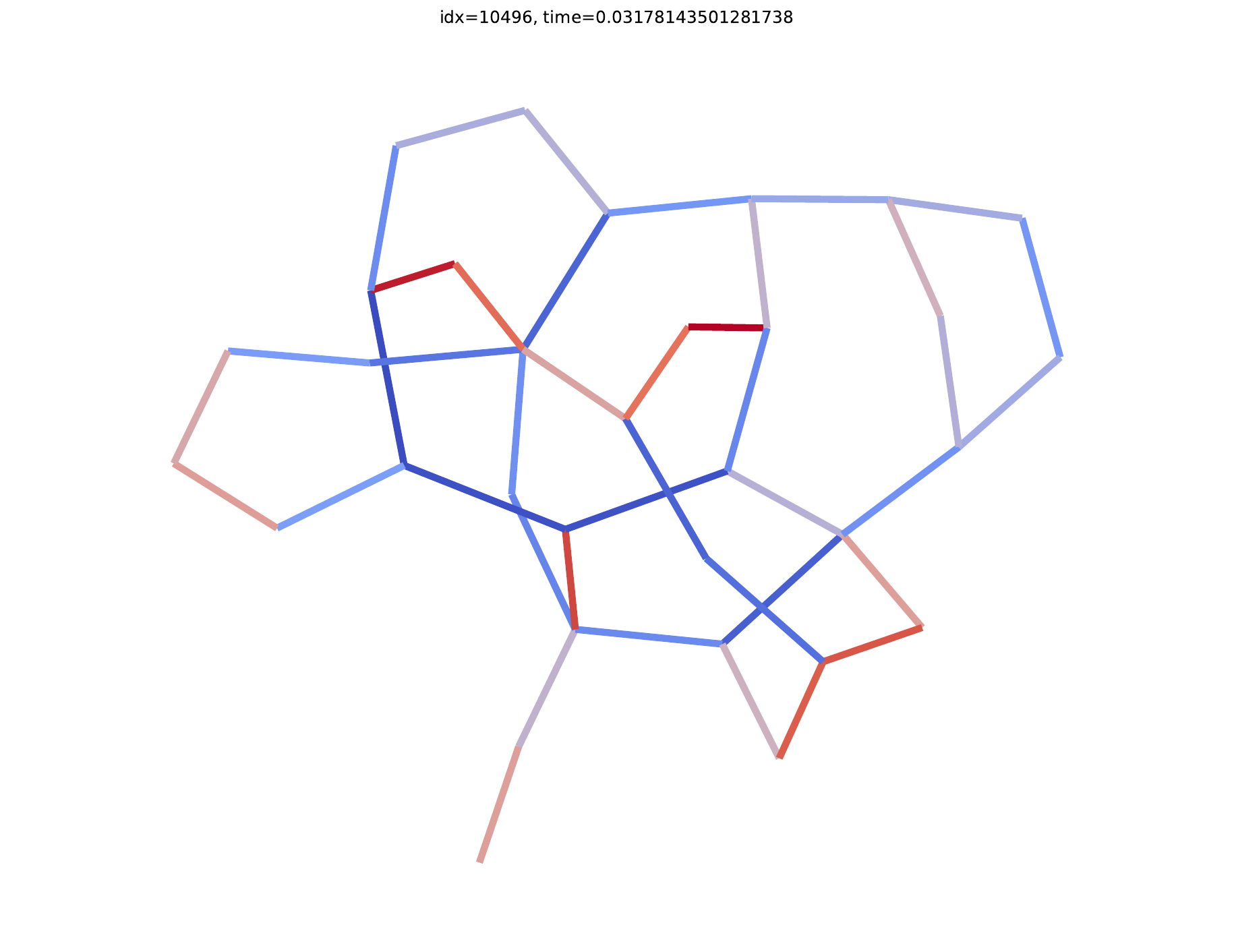} &
\imgcell{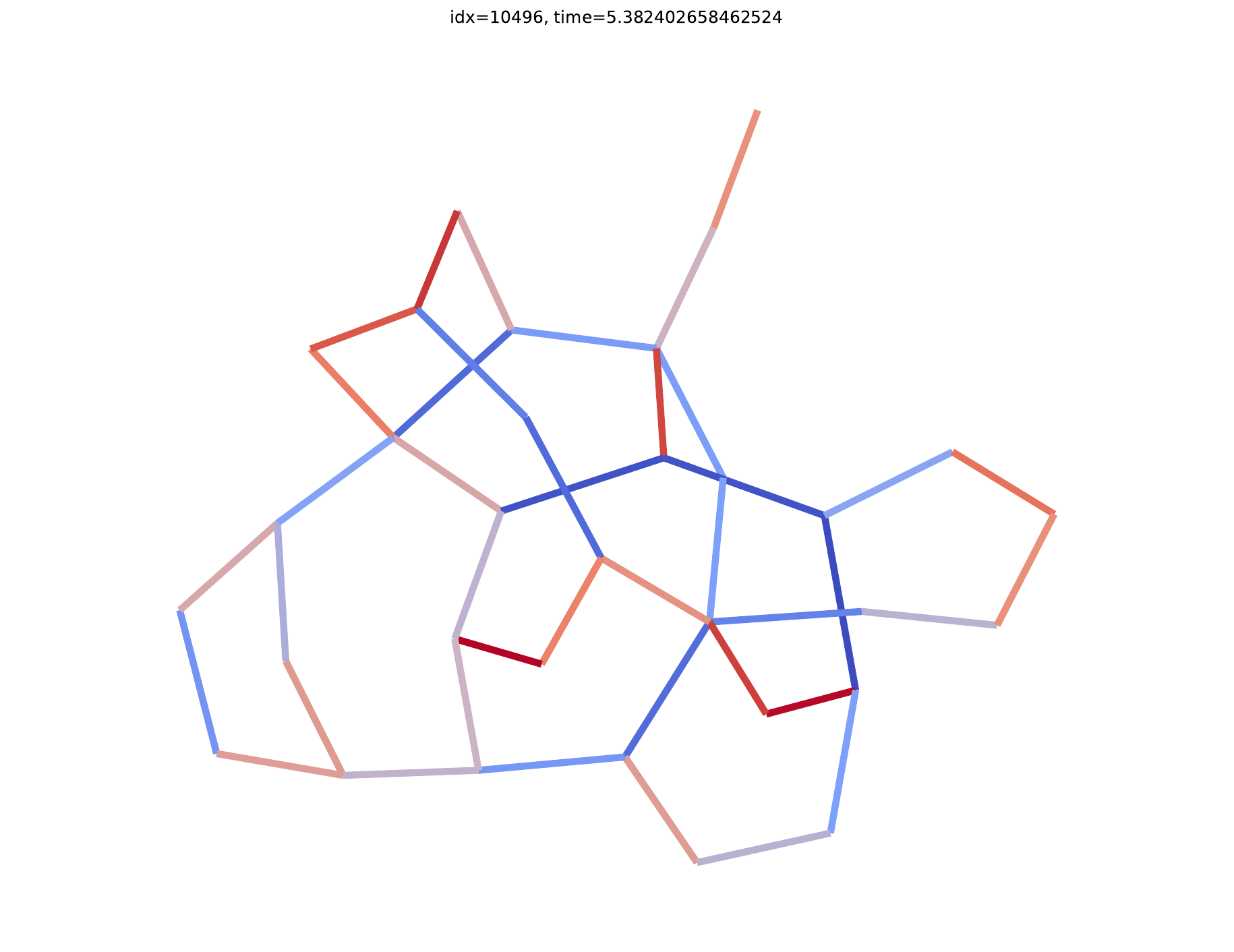} &
\imgcell{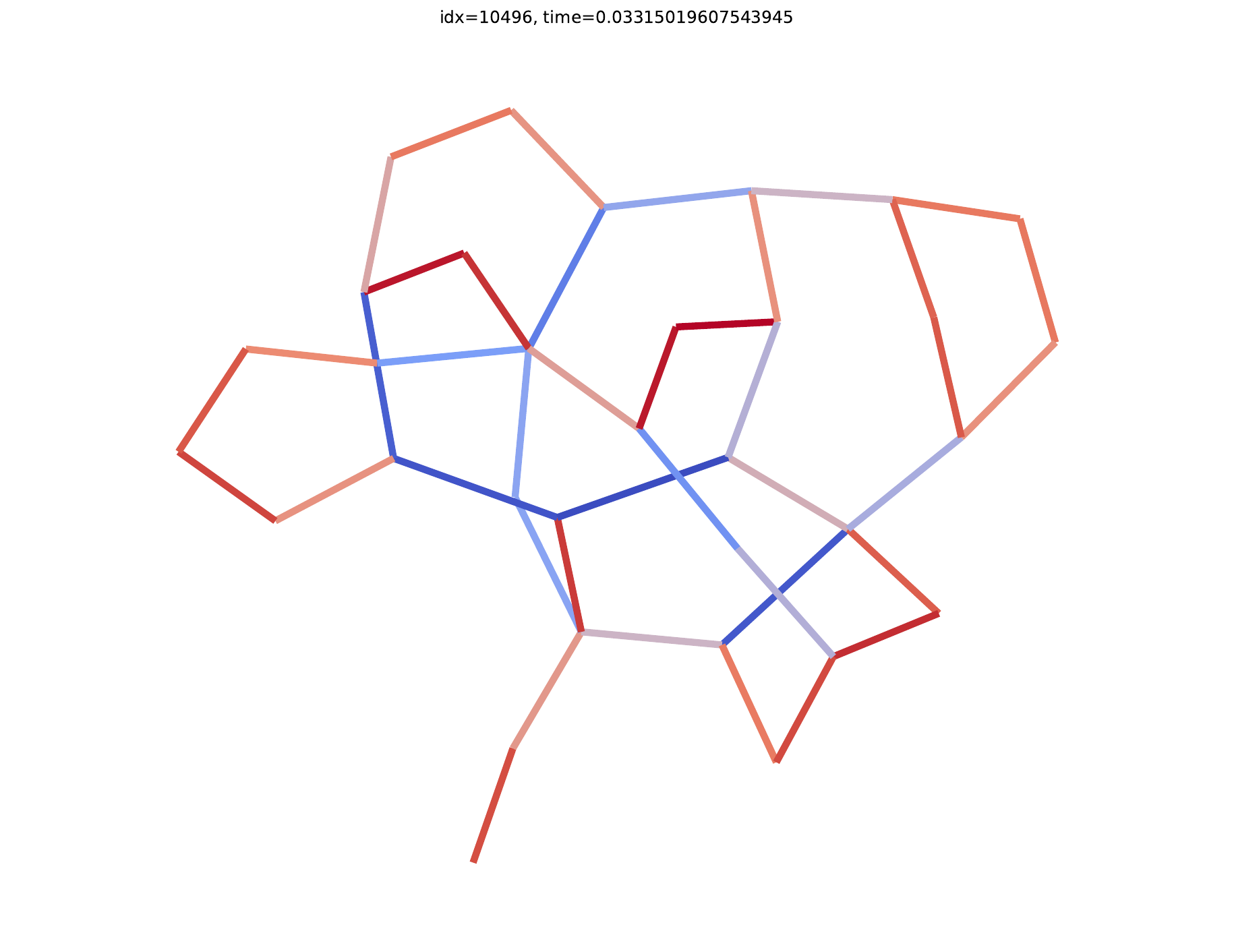} \\

&
t = 0.00s &
t = 0.32s &
t = 0.08s &
t = 0.05s &
t = 104.88s &
t = 0.04s &
t = 0.03s &
t = 0.04s &
t = 0.05s &
t = 0.03s &
t = 0.03s &
t = 0.03s \\

\makecell{\bfseries plskz362\\N = 362\\M = 880} &
\imgcell{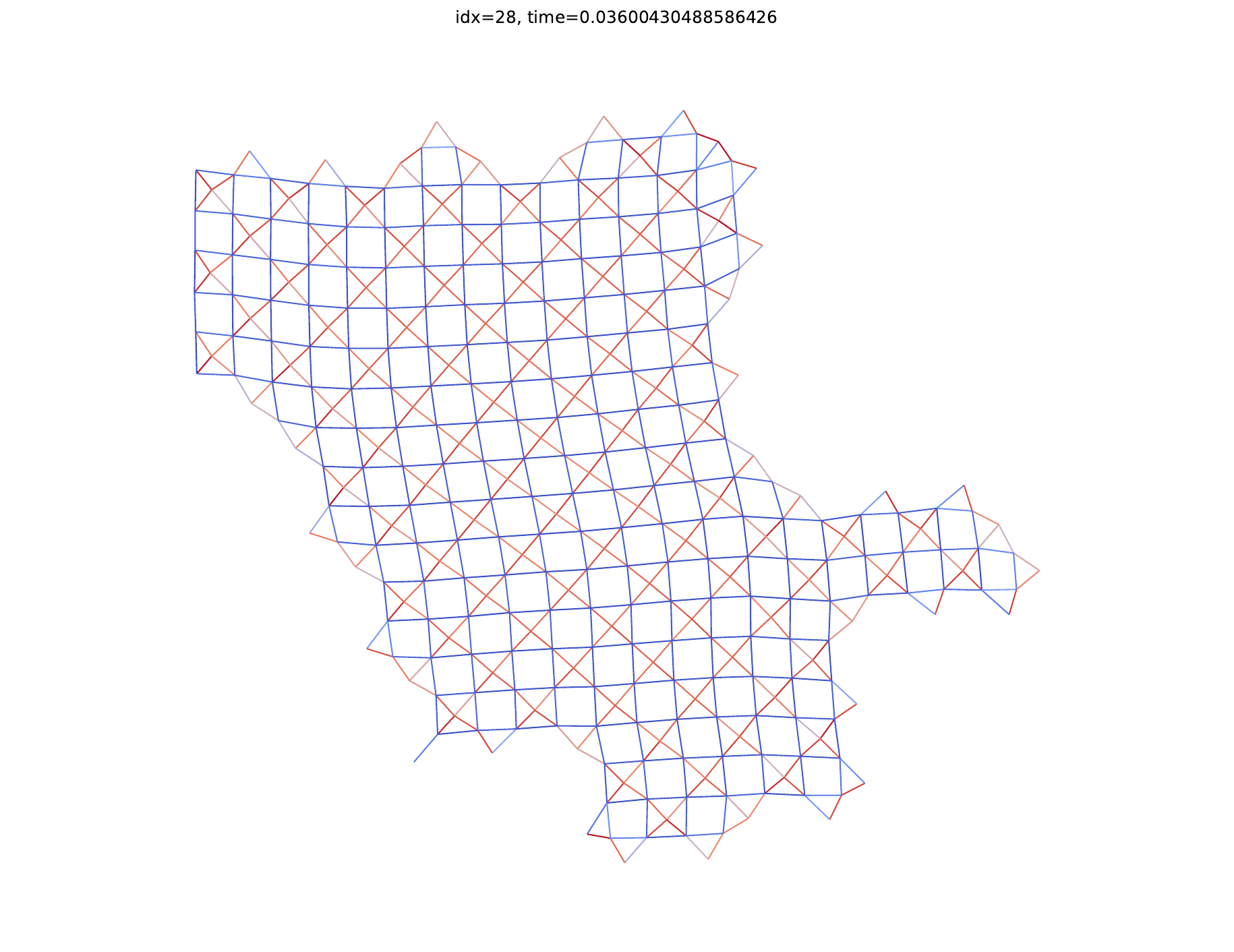} &
\imgcell{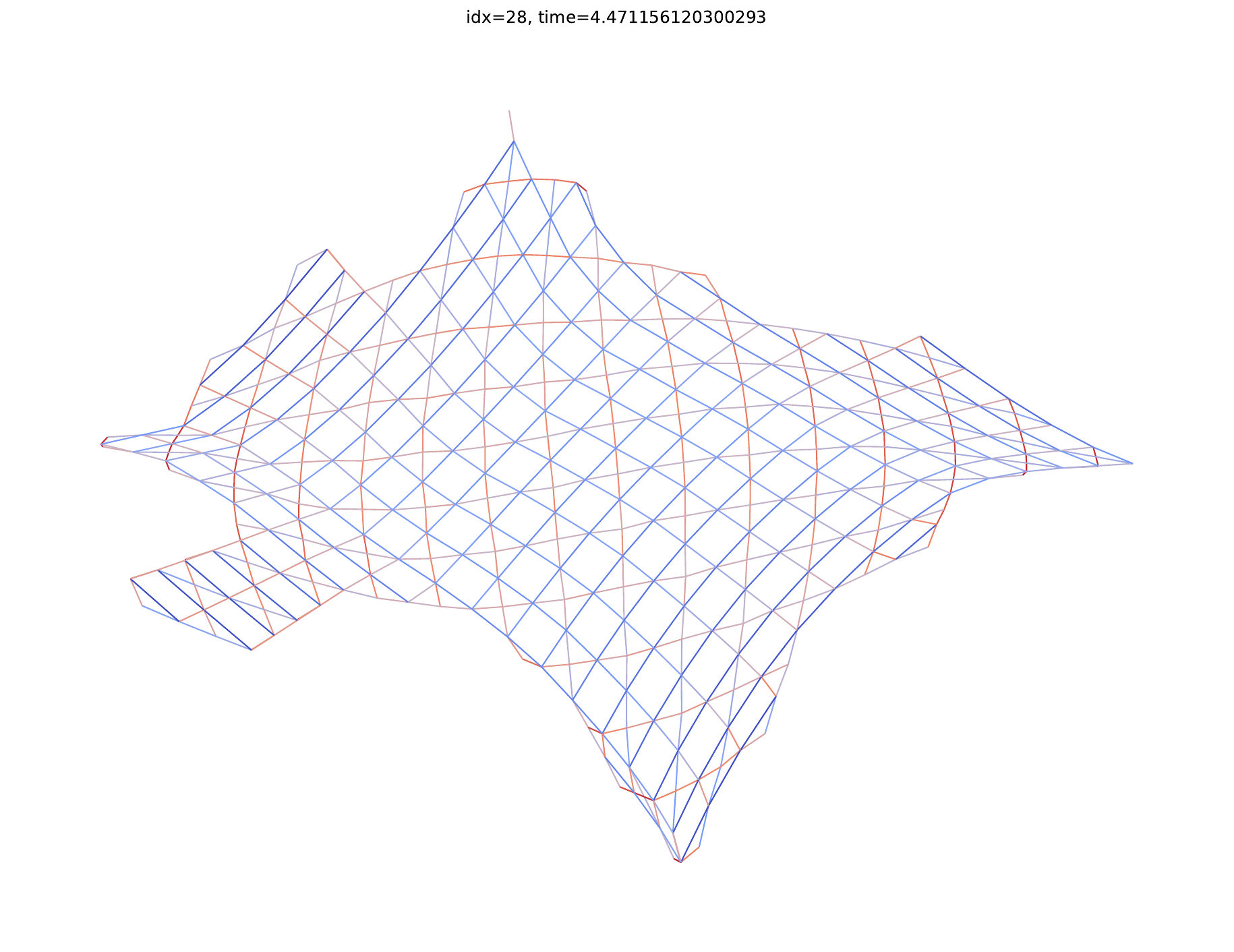} &
\imgcell{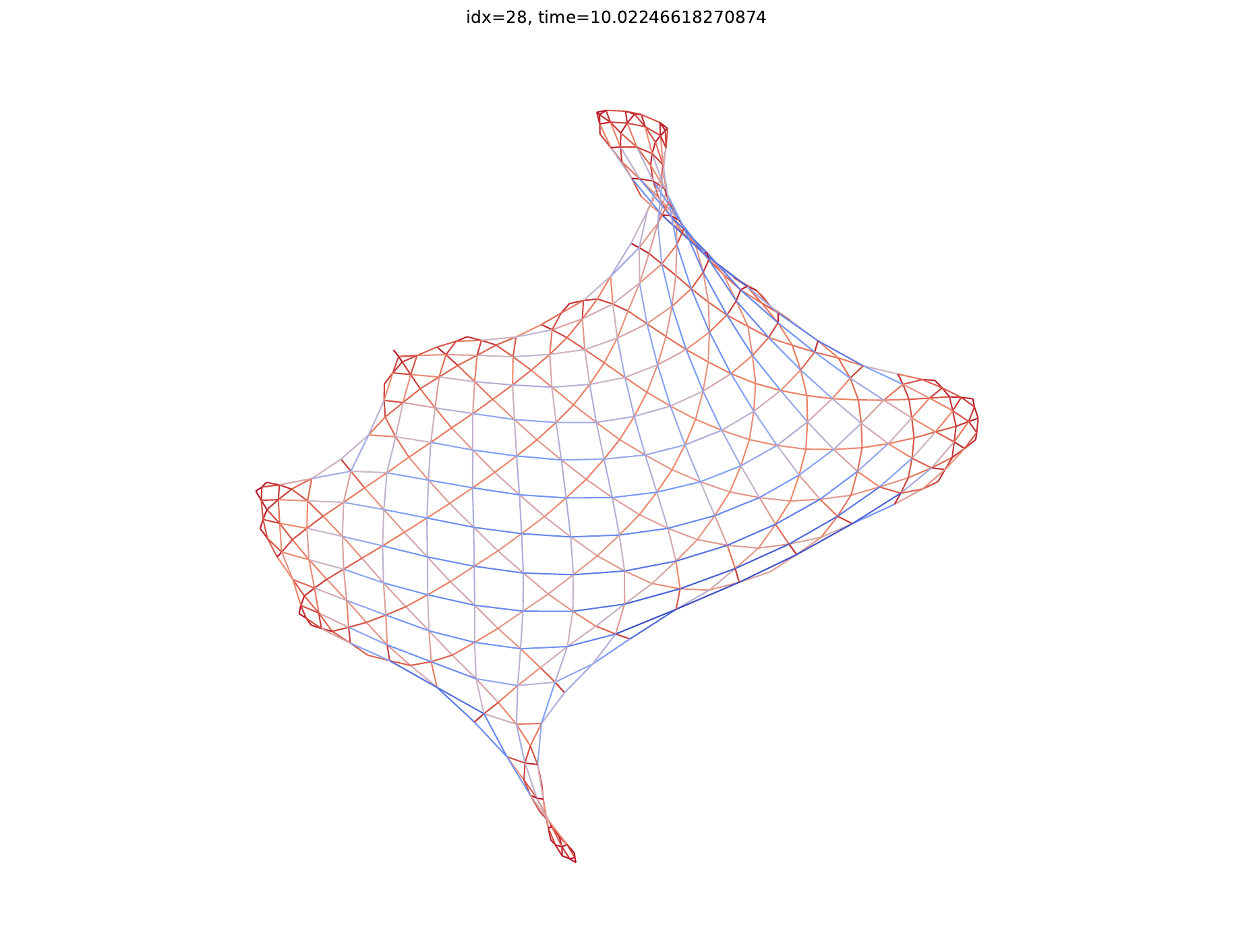} &
\imgcell{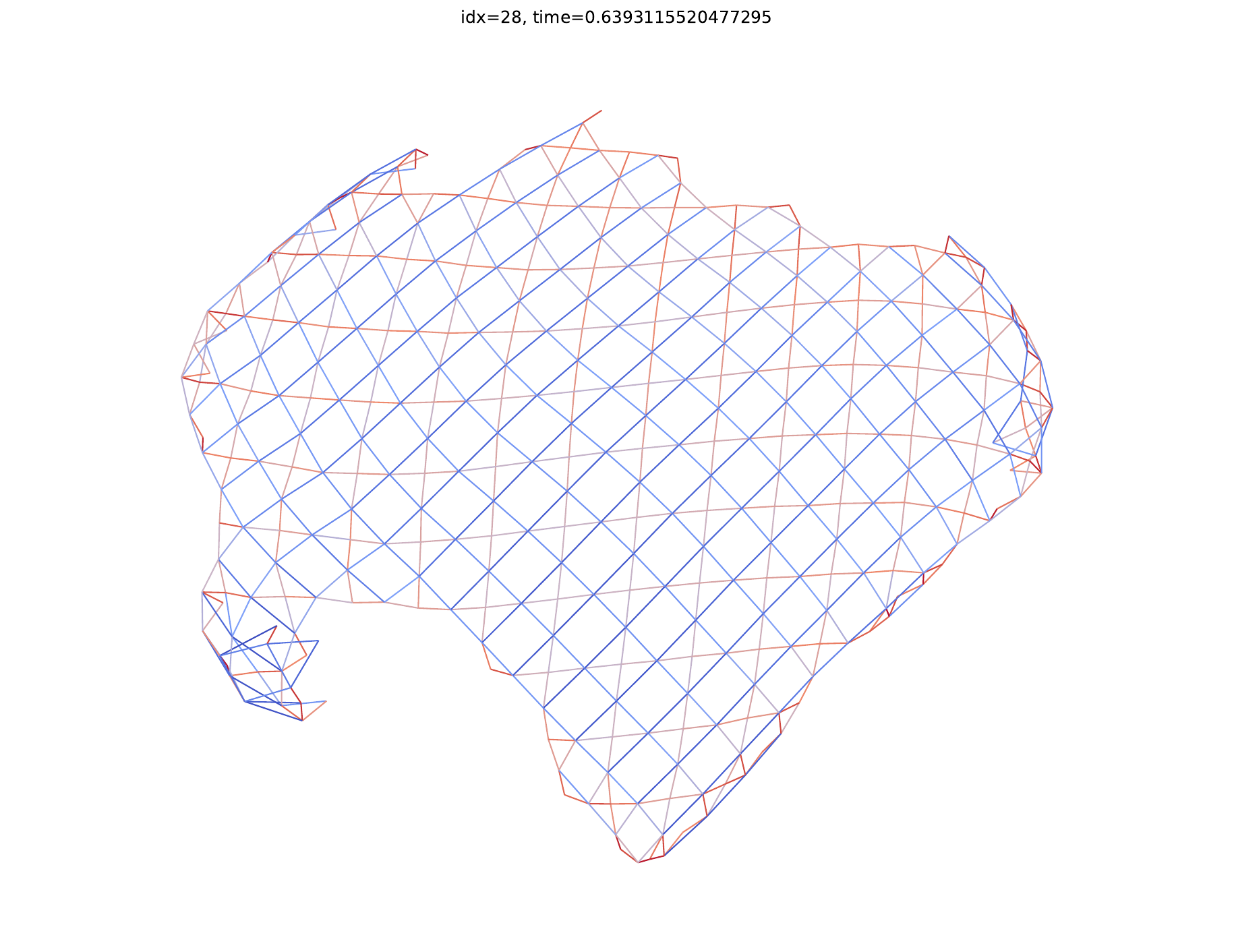} &
\imgcell{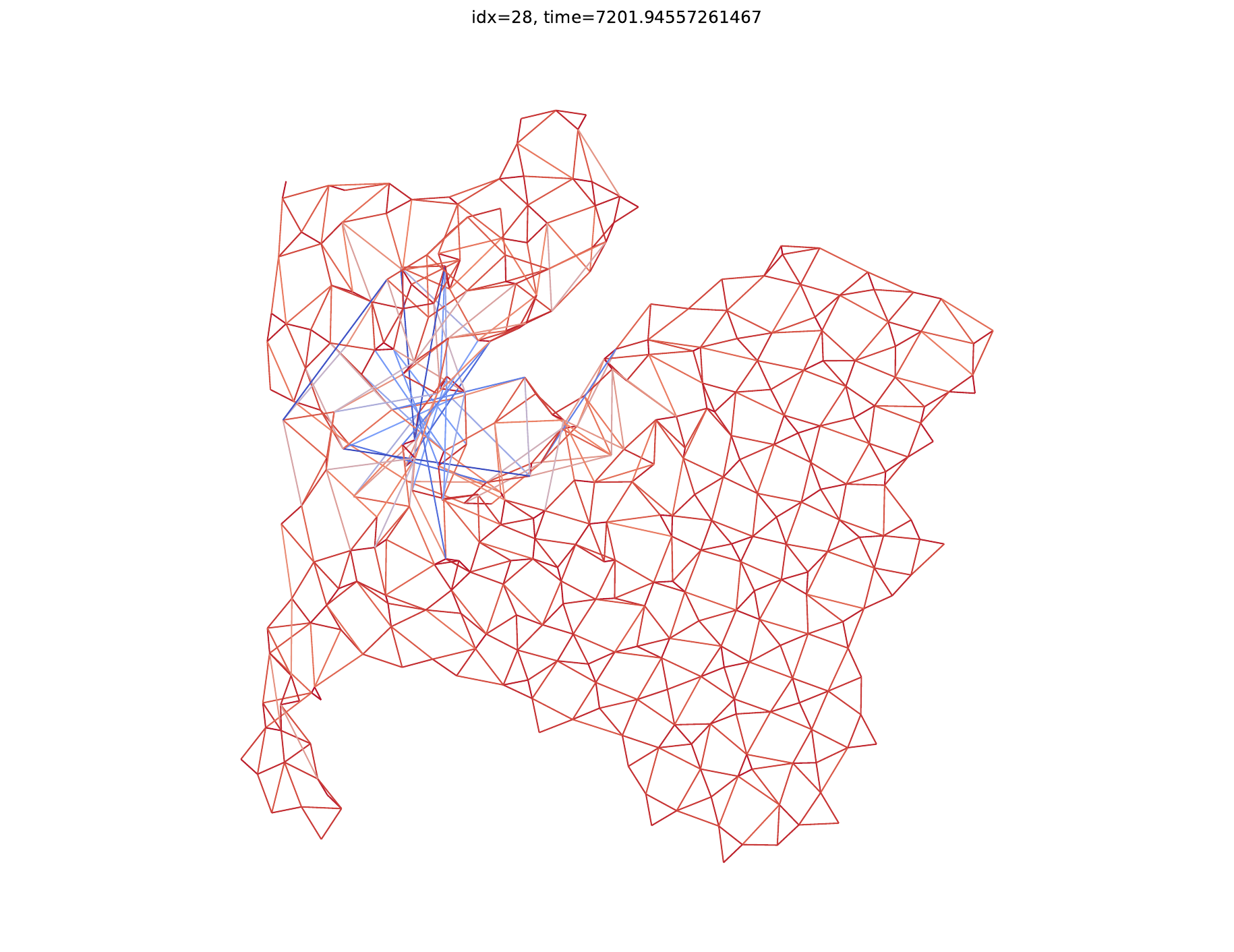} &
\imgcell{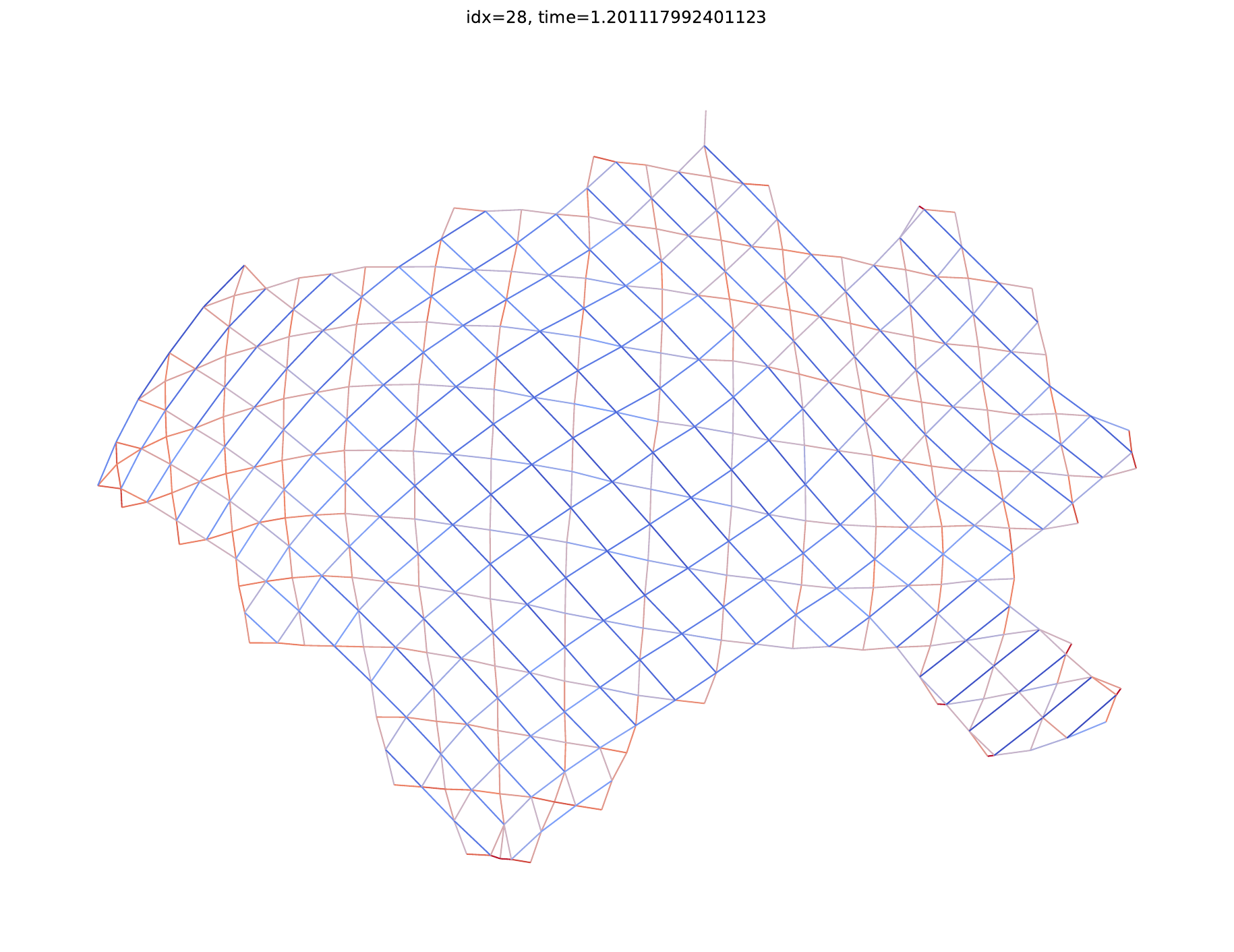} &
\imgcell{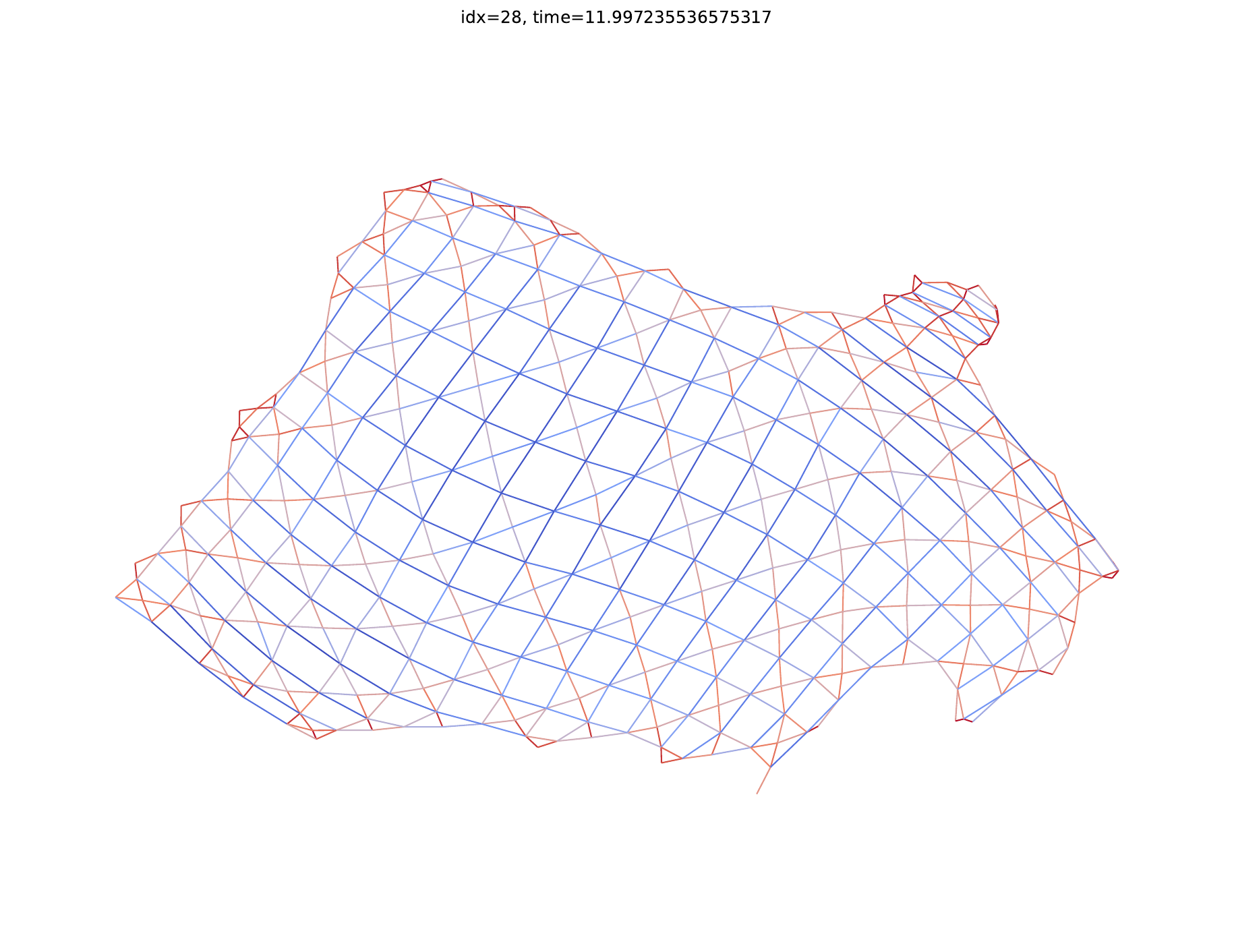} &
\imgcell{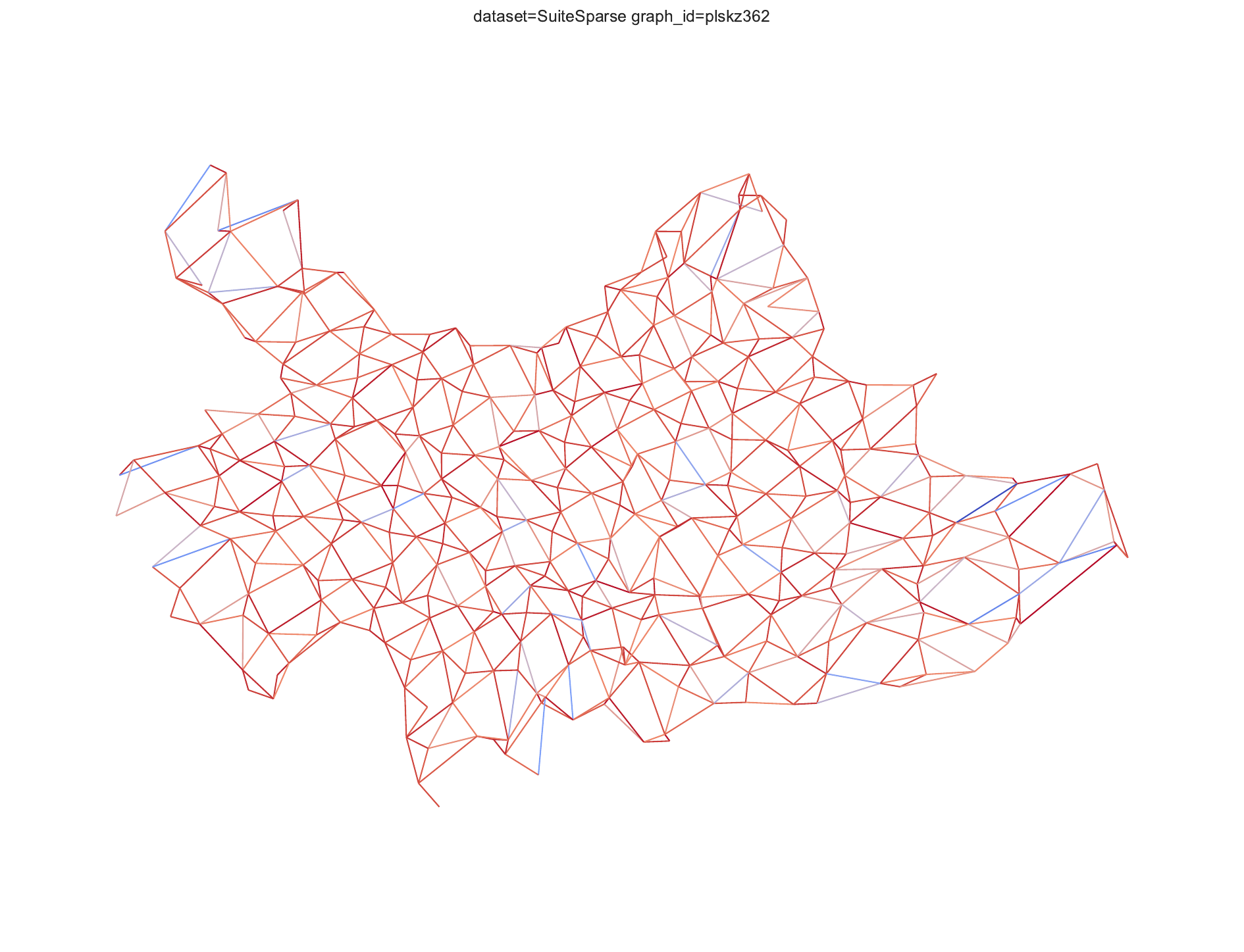} &
\imgcell{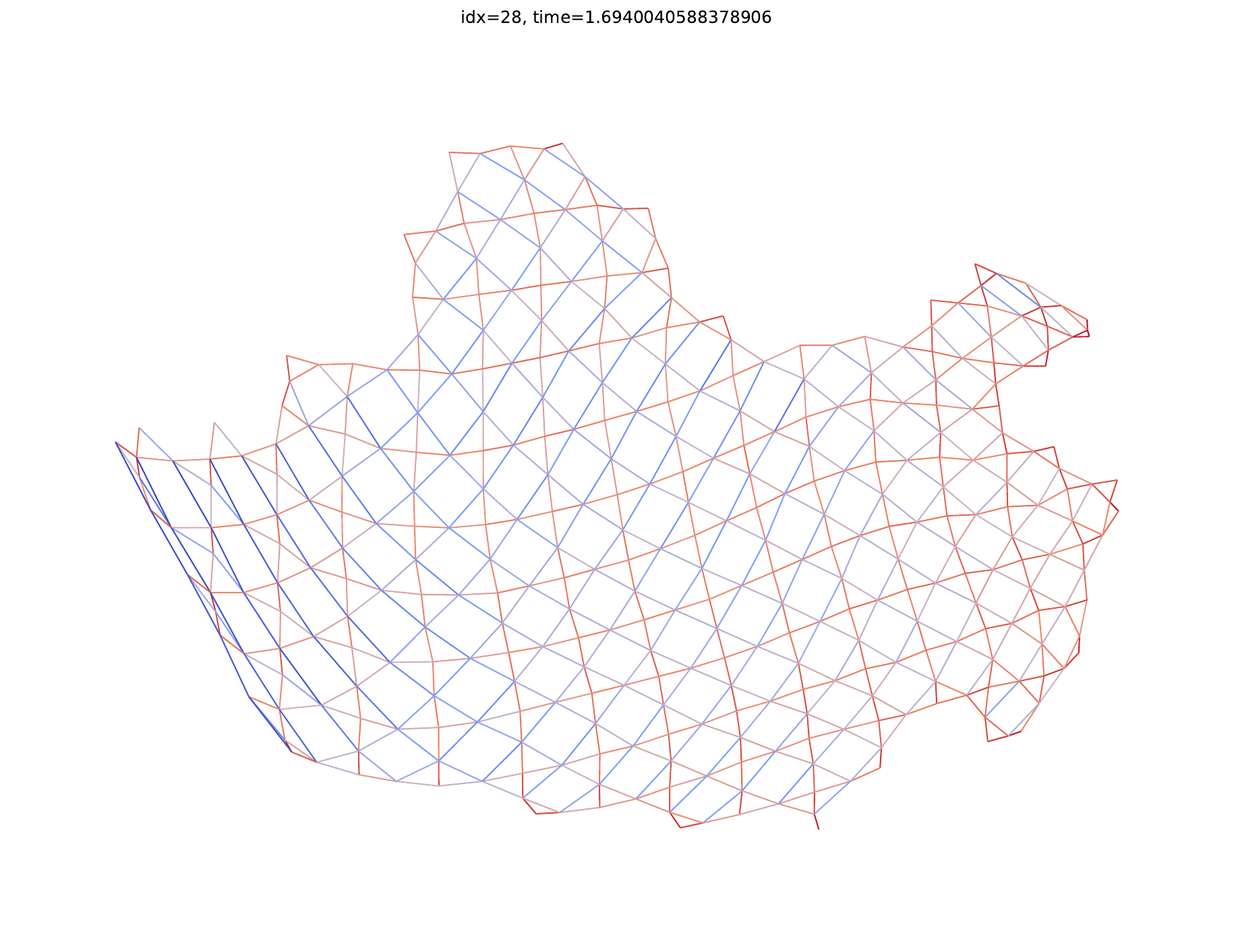} &
\imgcell{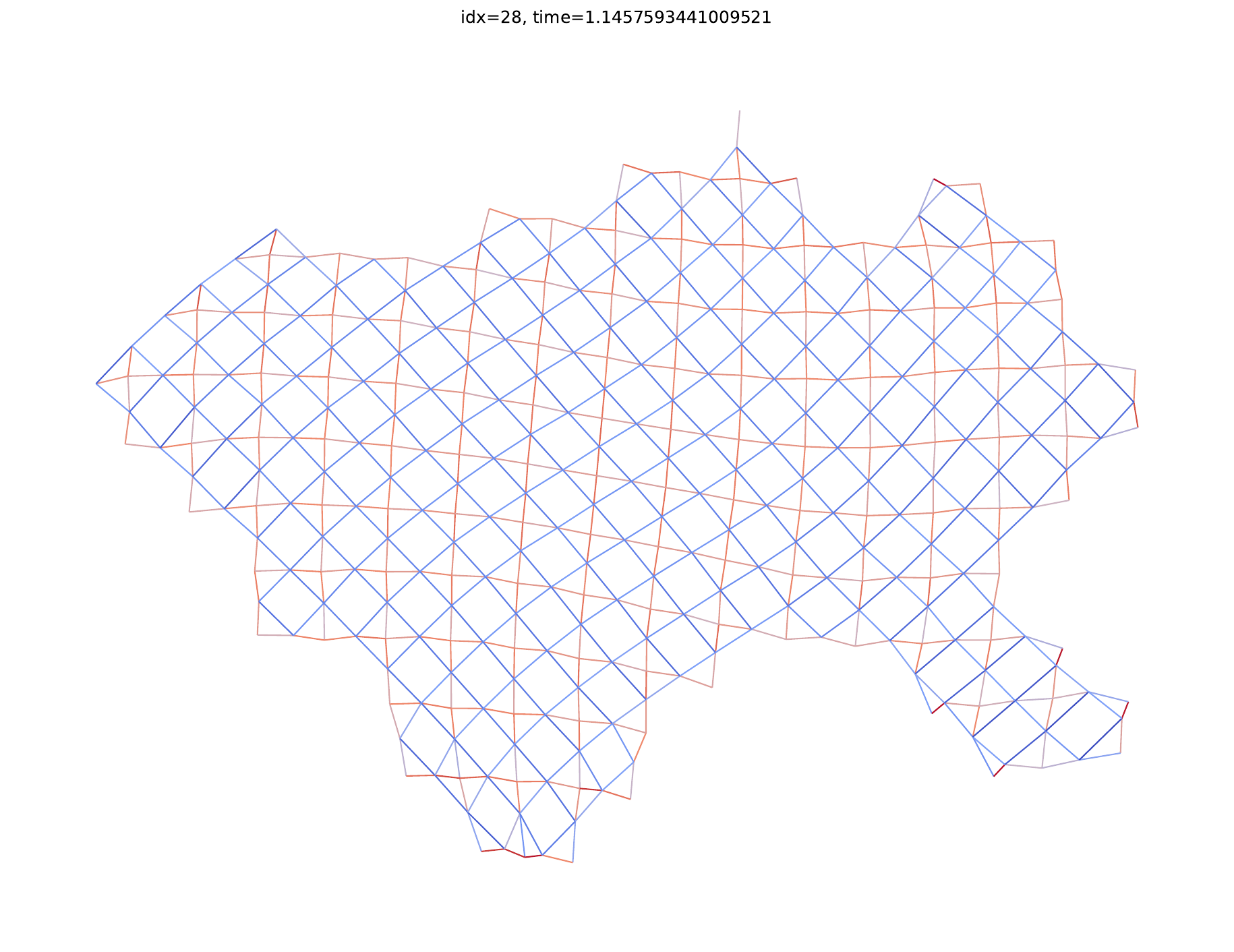} &
\imgcell{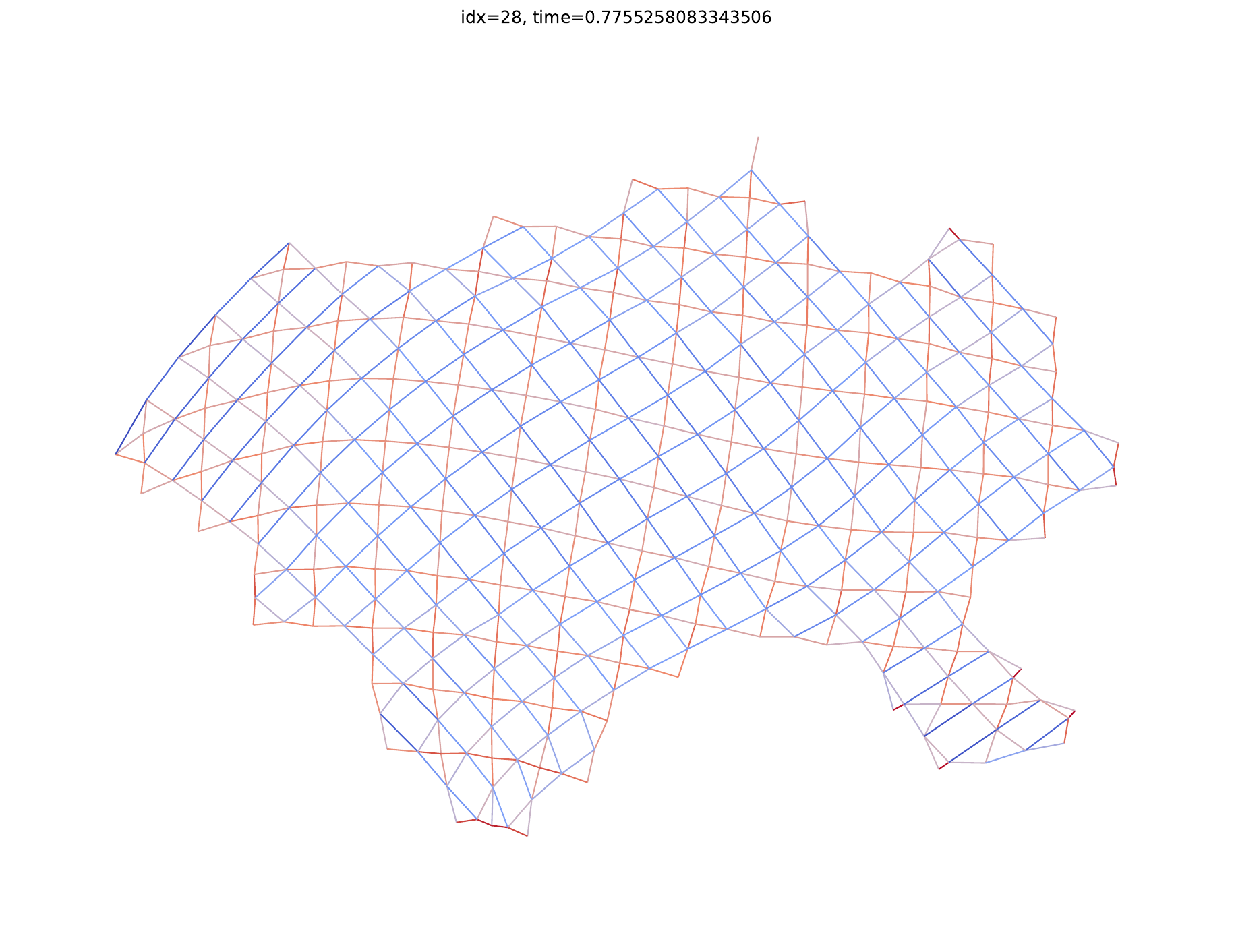} &
\imgcell{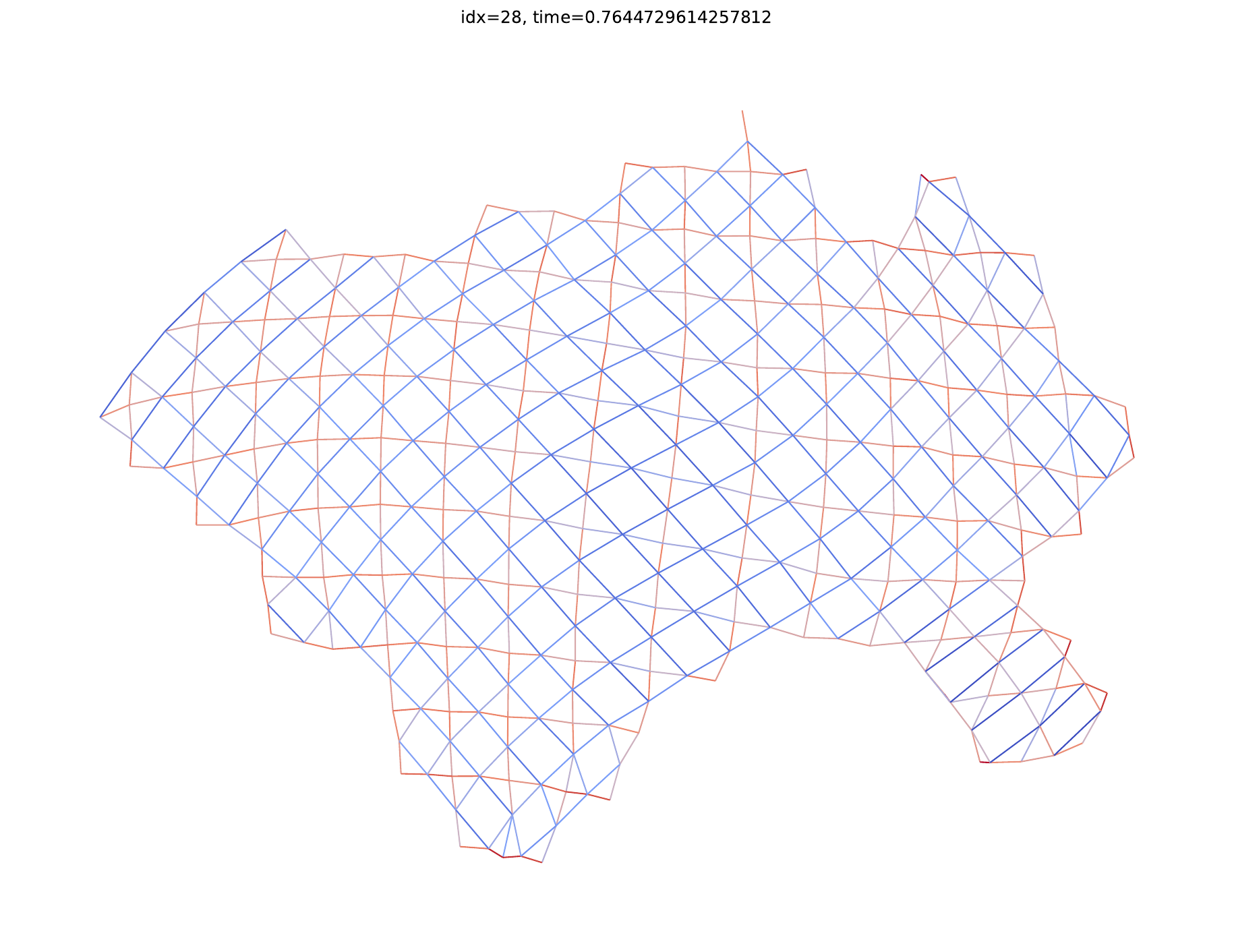} \\

&
t = 0.04s &
t = 4.47s &
t = 10.02s &
t = 0.64s &
t = 7200.00s &
t = 0.60s &
t = 0.49s &
t = 0.44s &
t = 0.58s &
t = 0.42s &
t = 0.65s &
t = 0.51s \\

\makecell{\bfseries bfwa782\\N = 782\\M = 3394} &
\imgcell{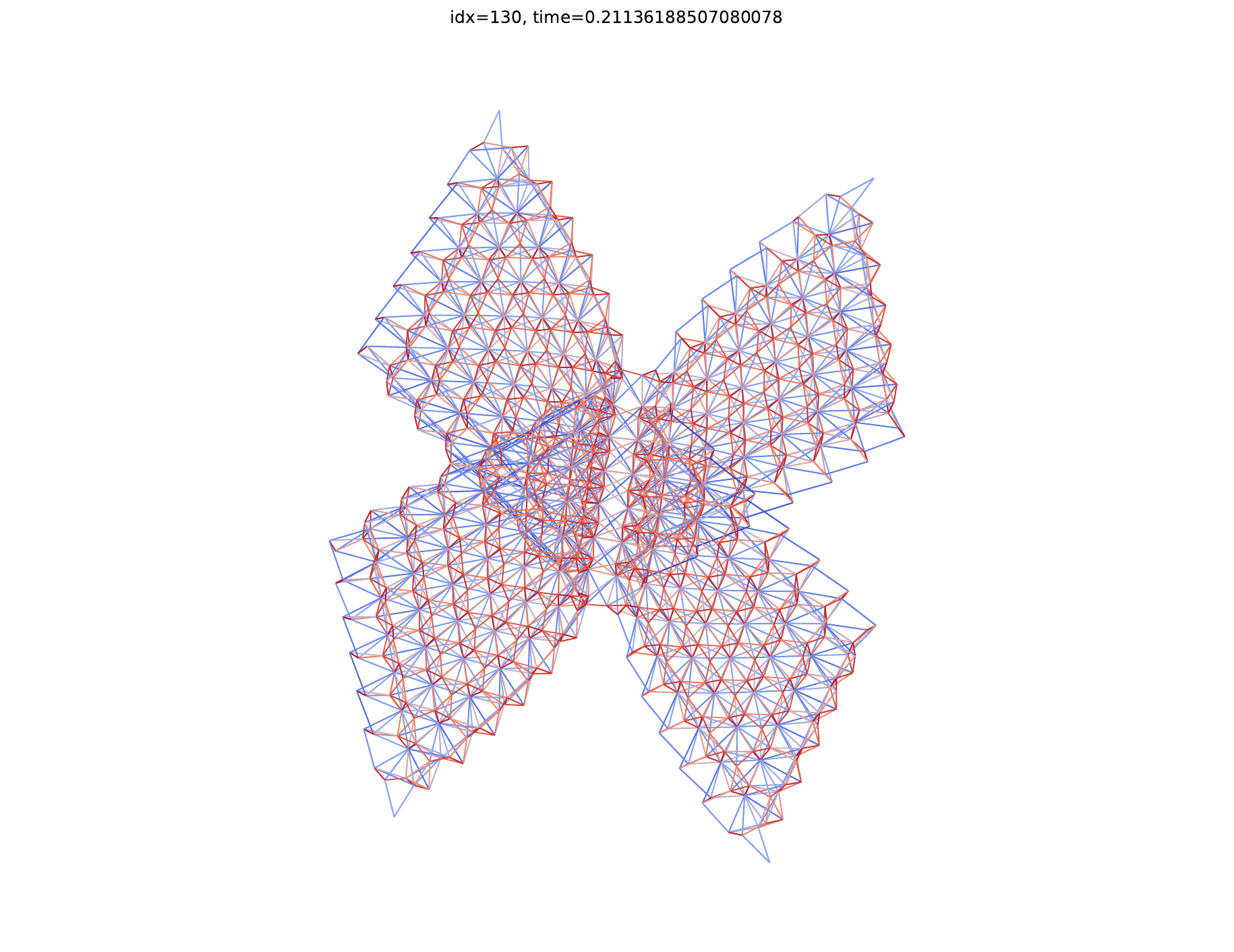} &
\imgcell{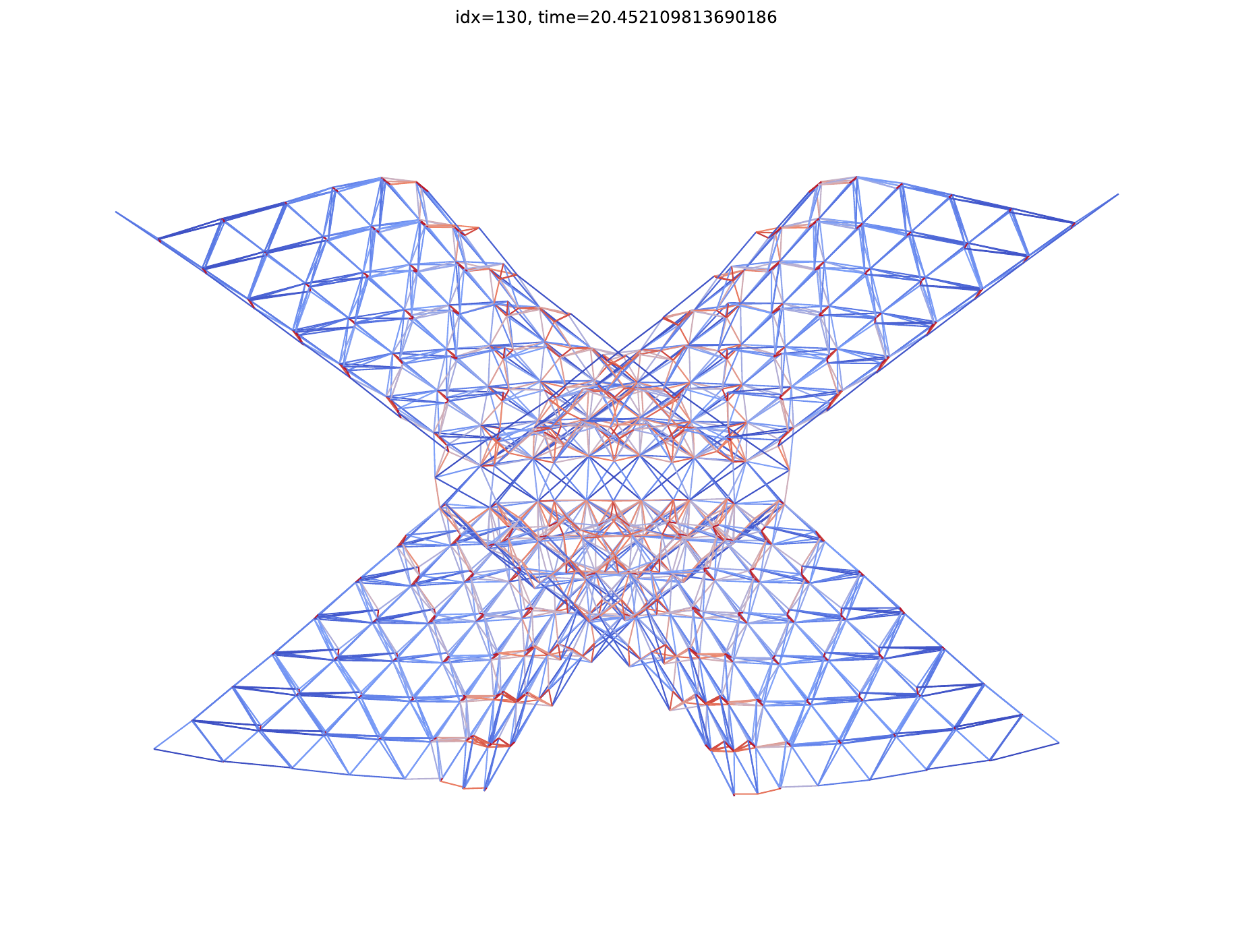} &
\imgcell{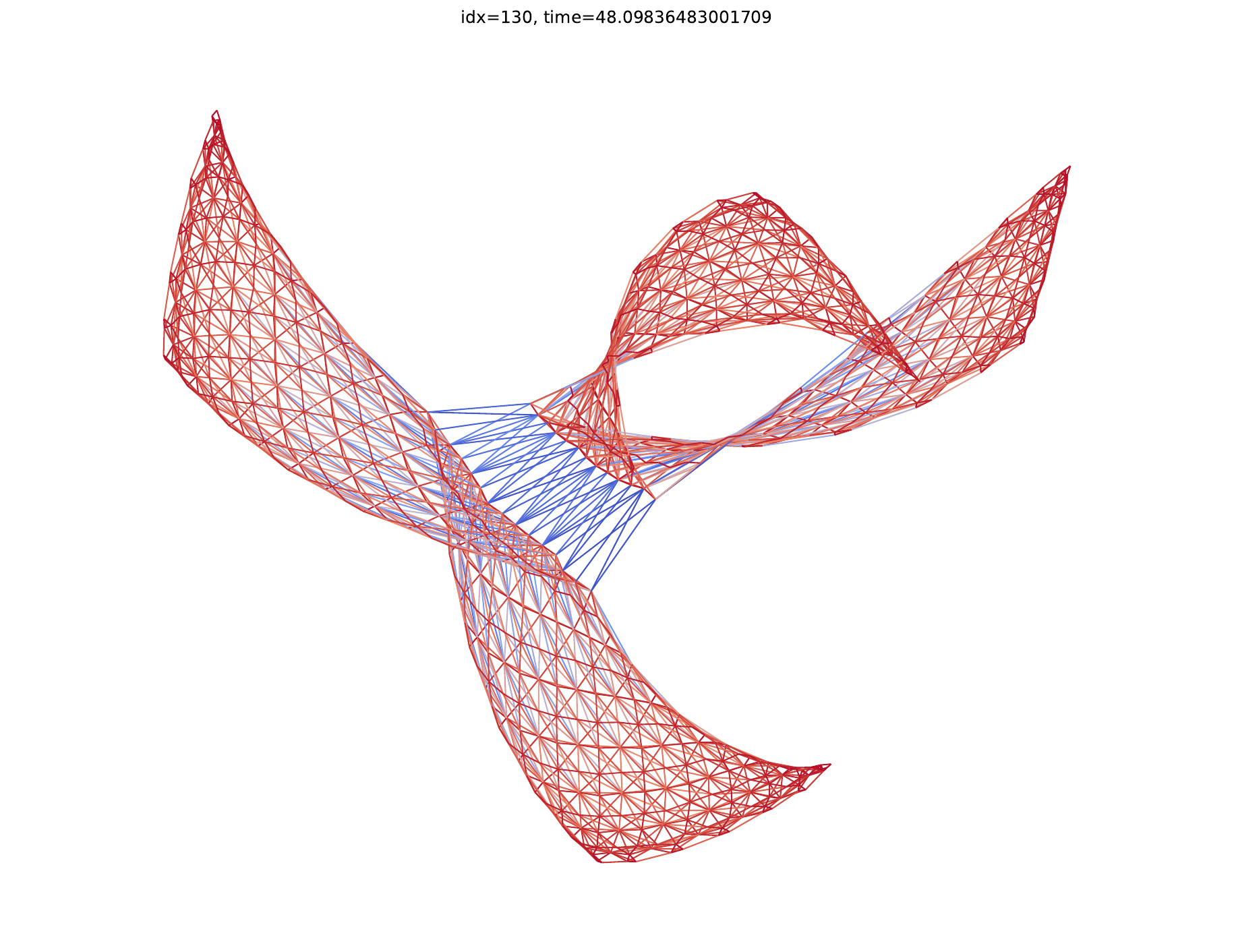} &
\imgcell{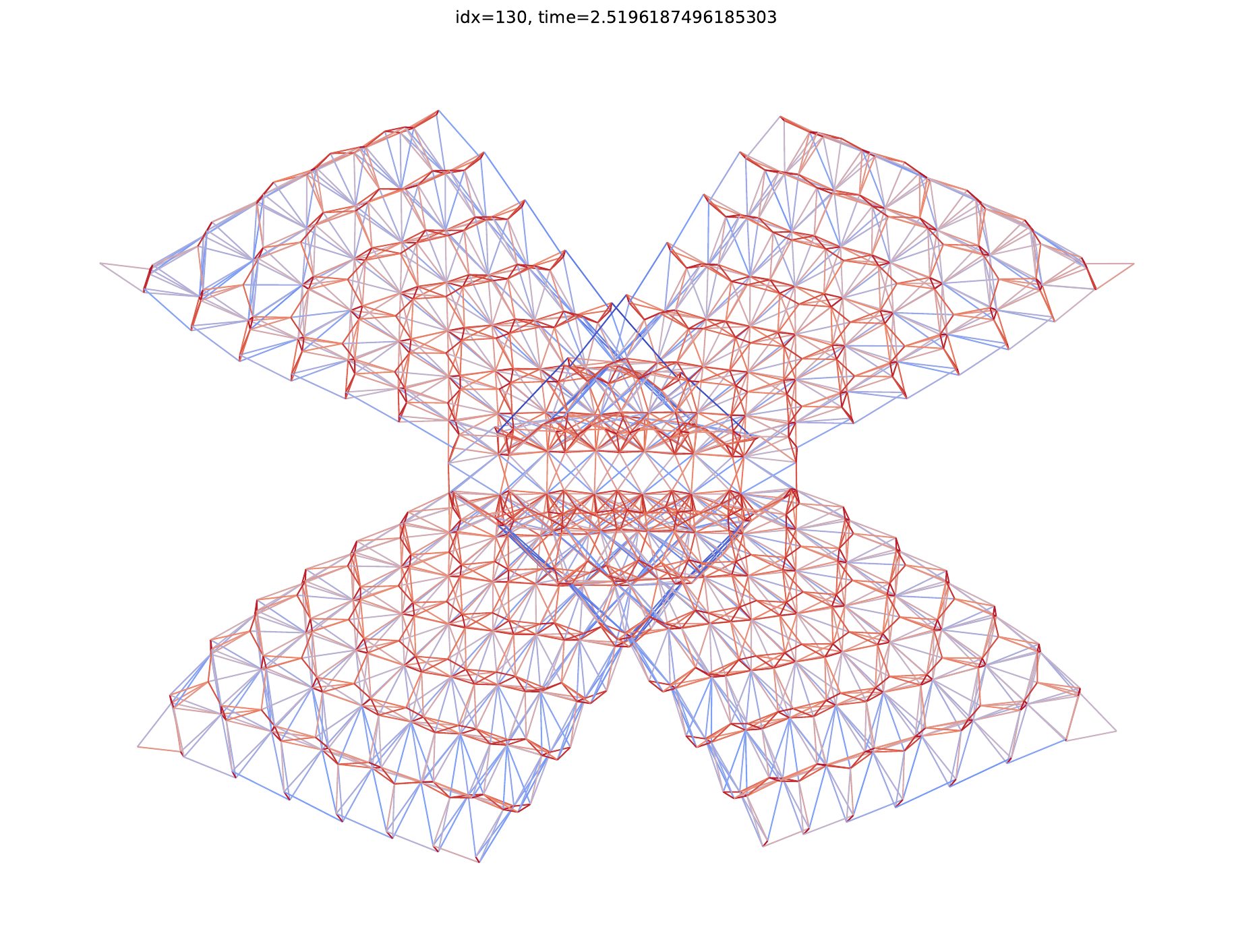} &
\imgcell{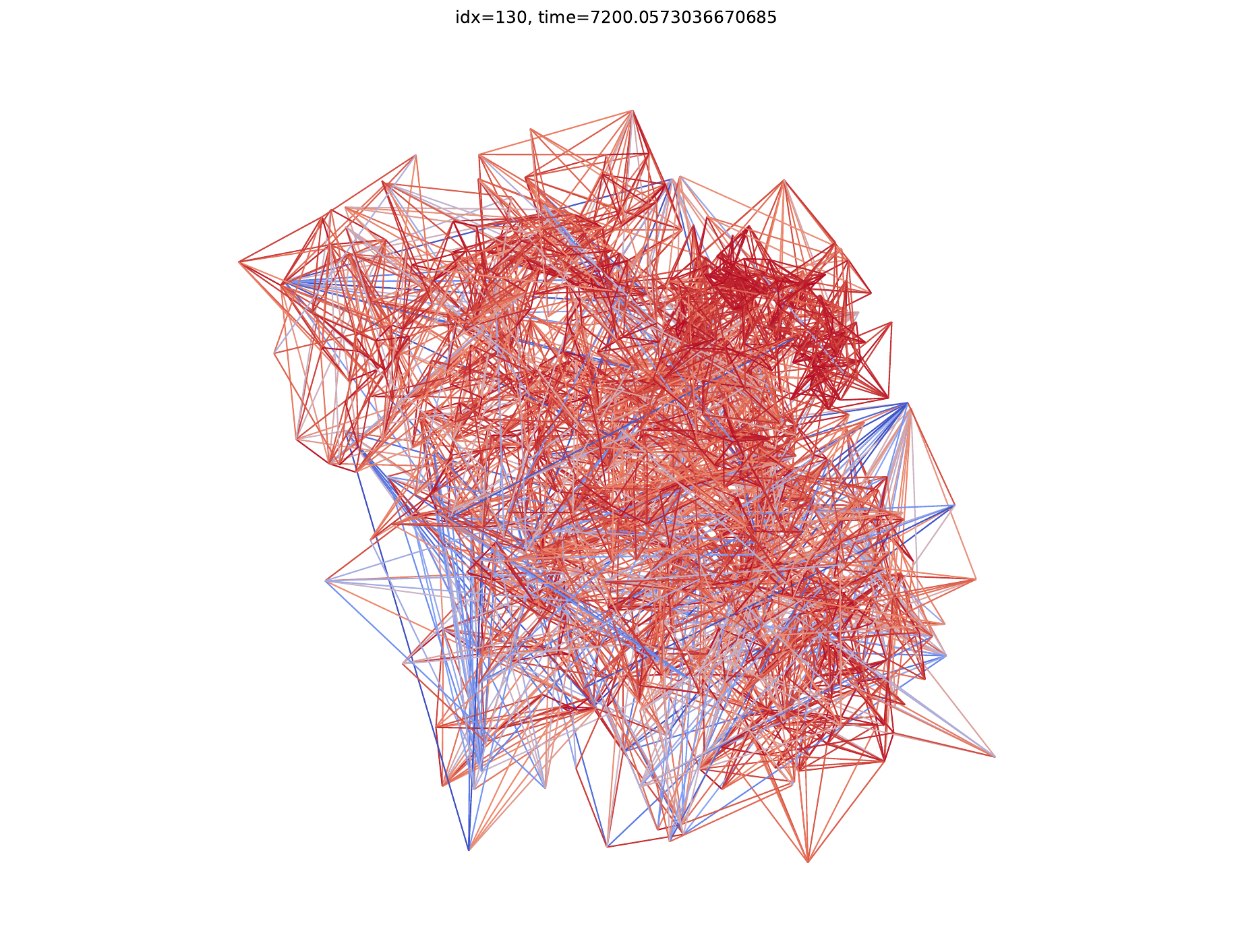} &
\imgcell{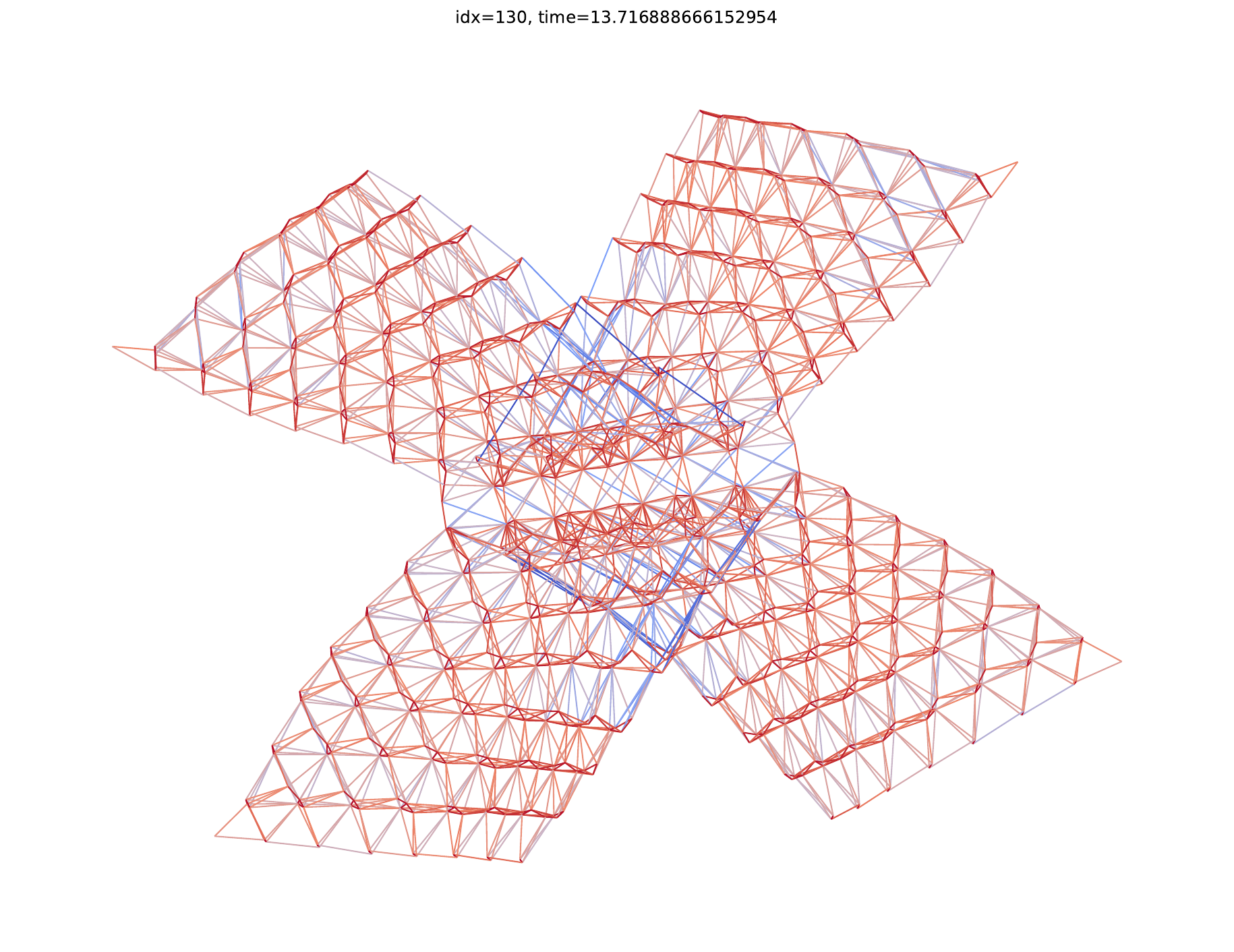} &
\imgcell{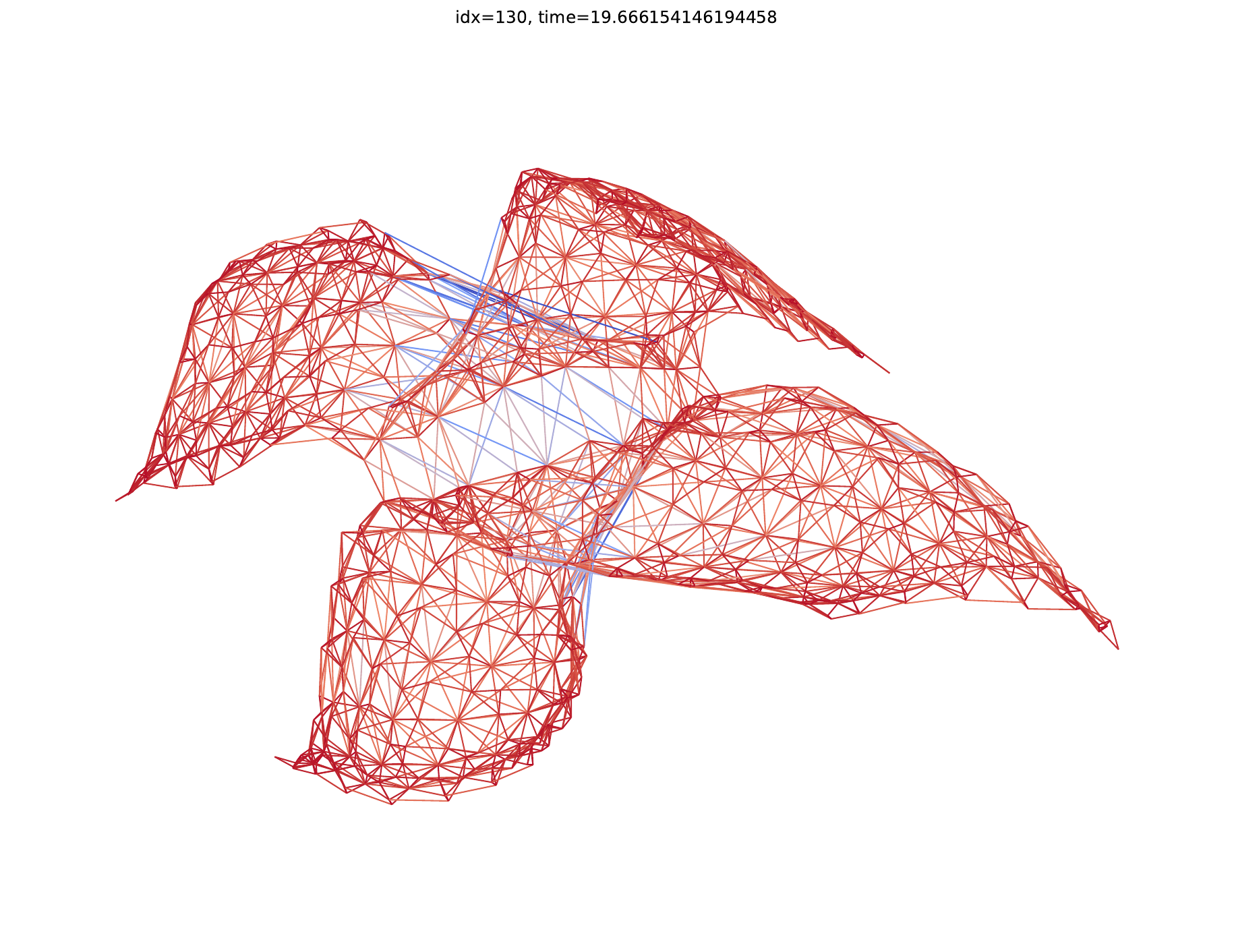} &
\imgcell{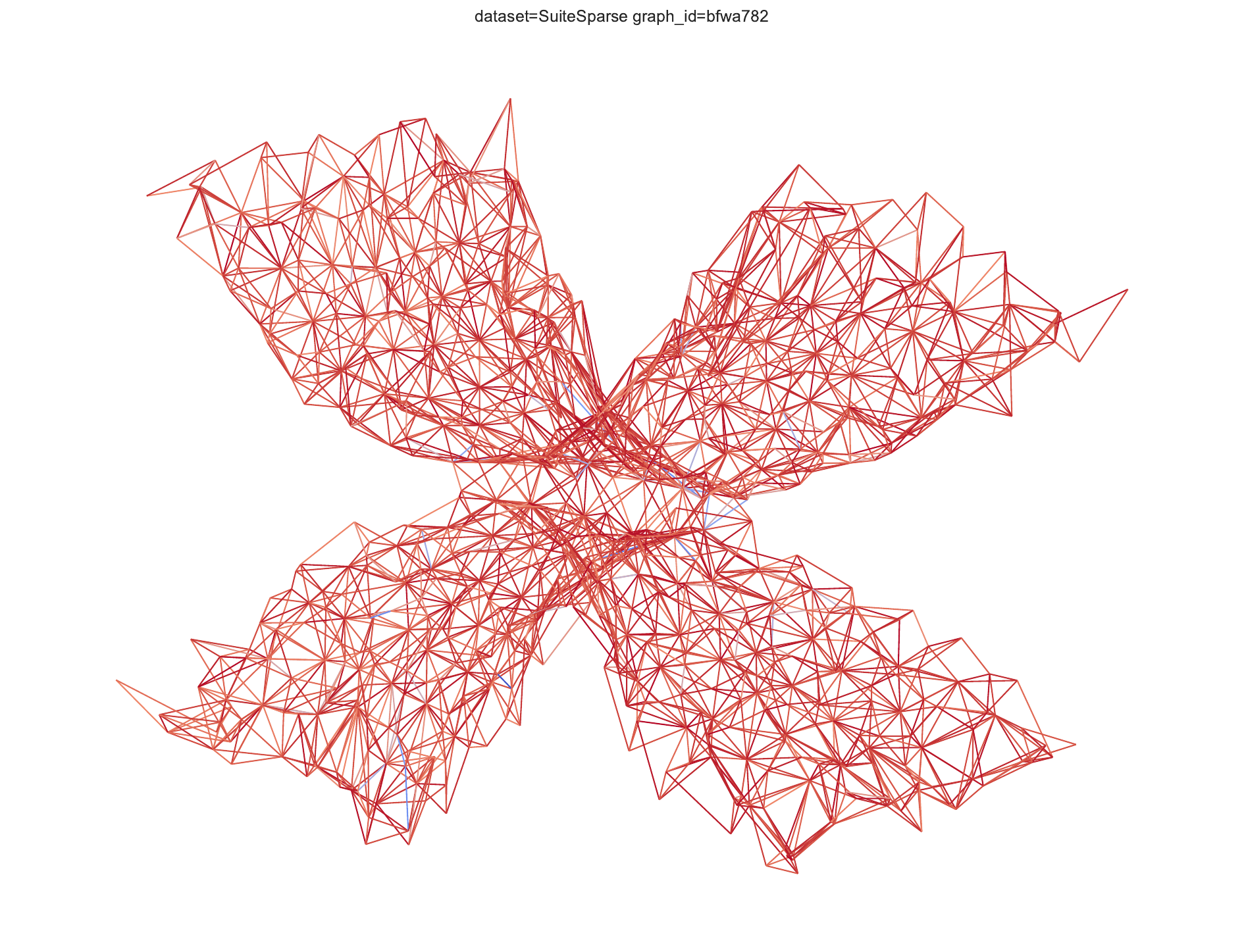} &
\imgcell{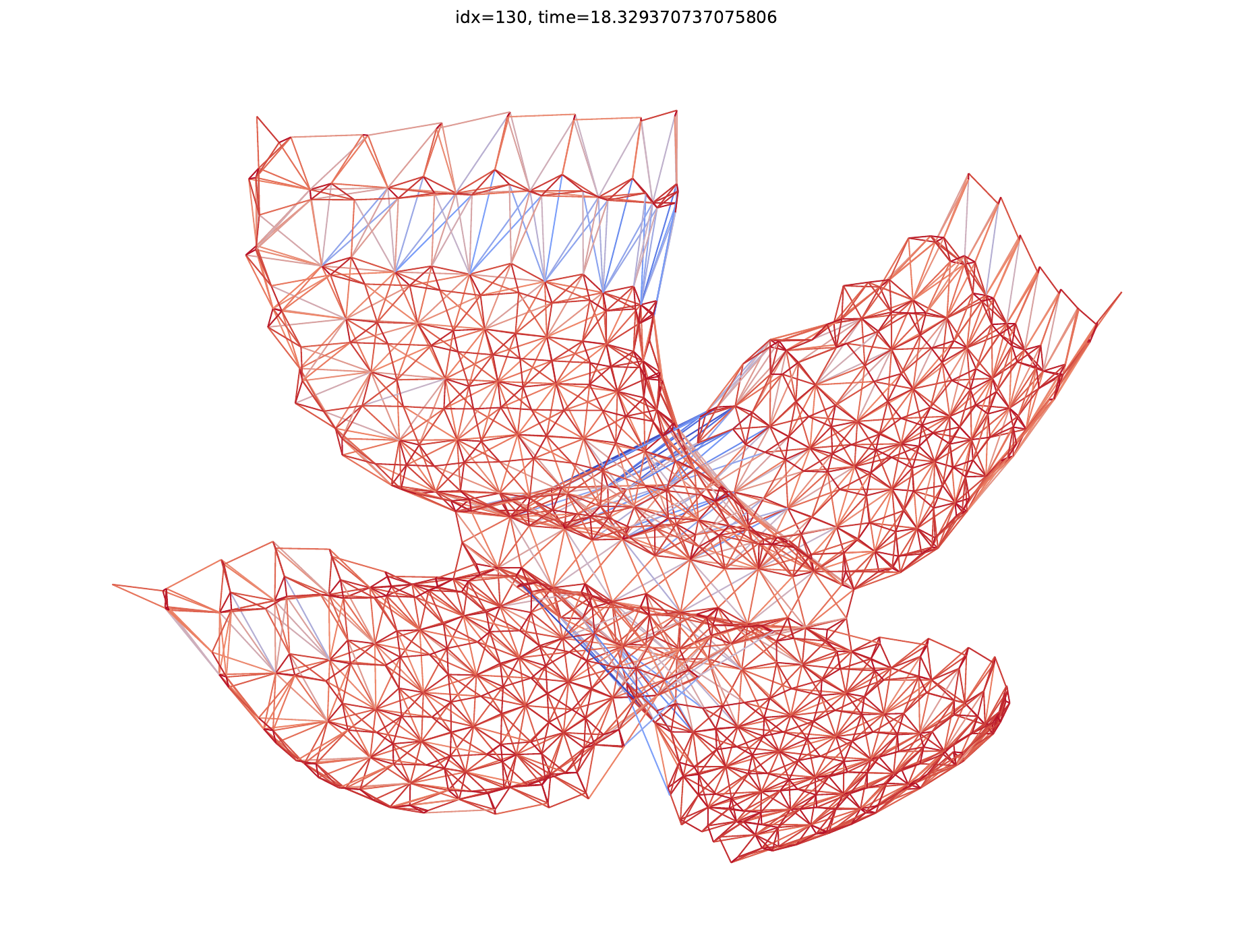} &
\imgcell{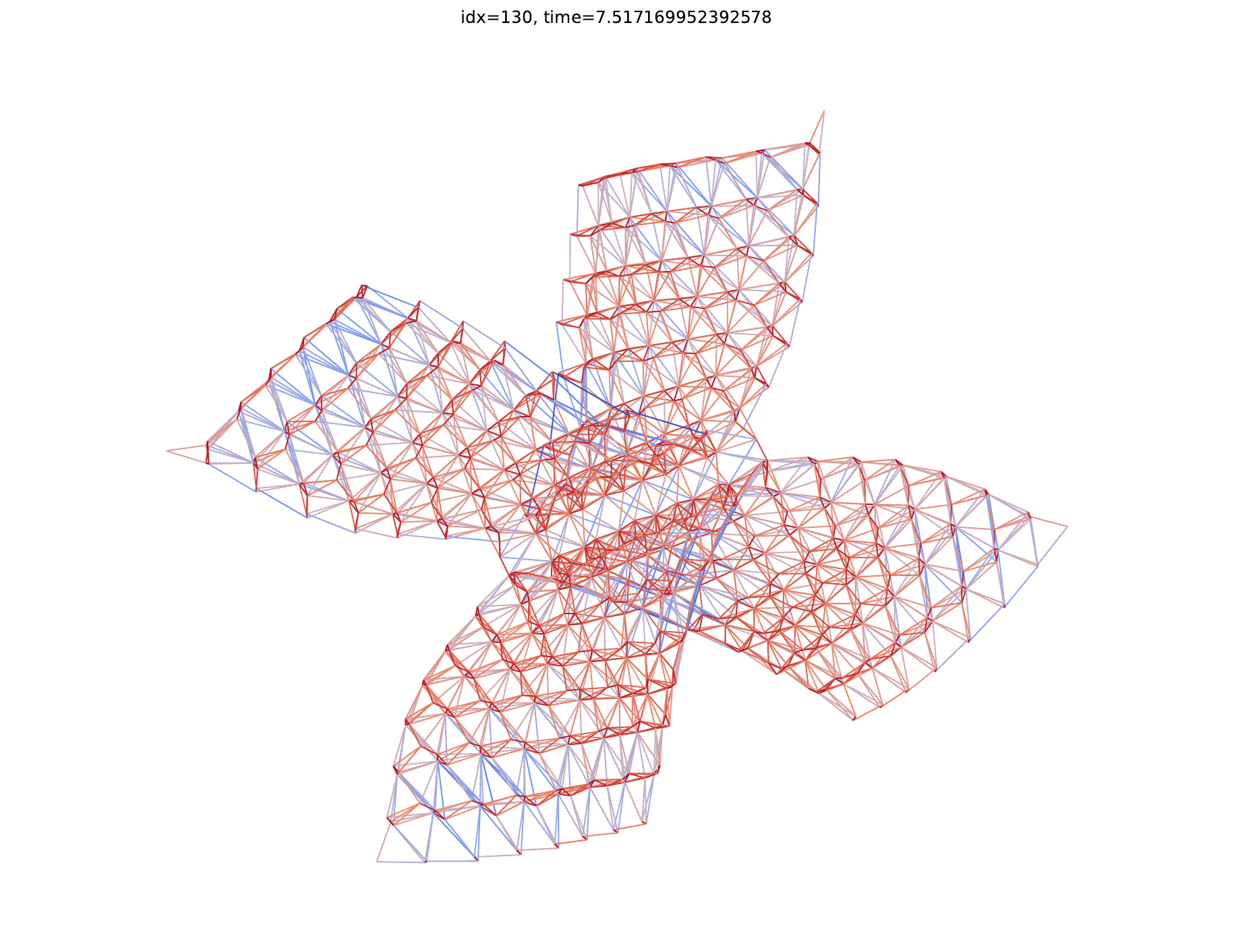} &
\imgcell{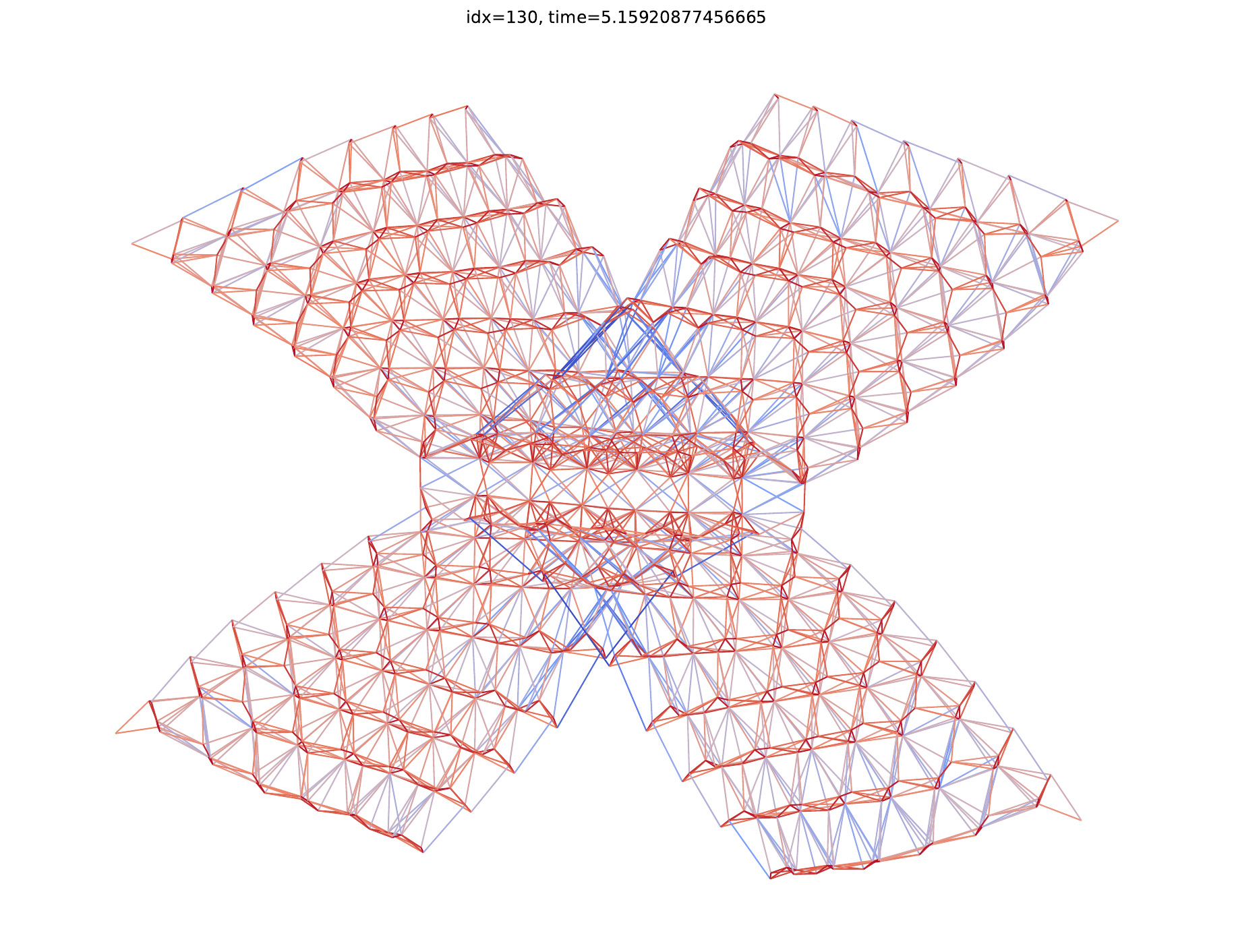} &
\imgcell{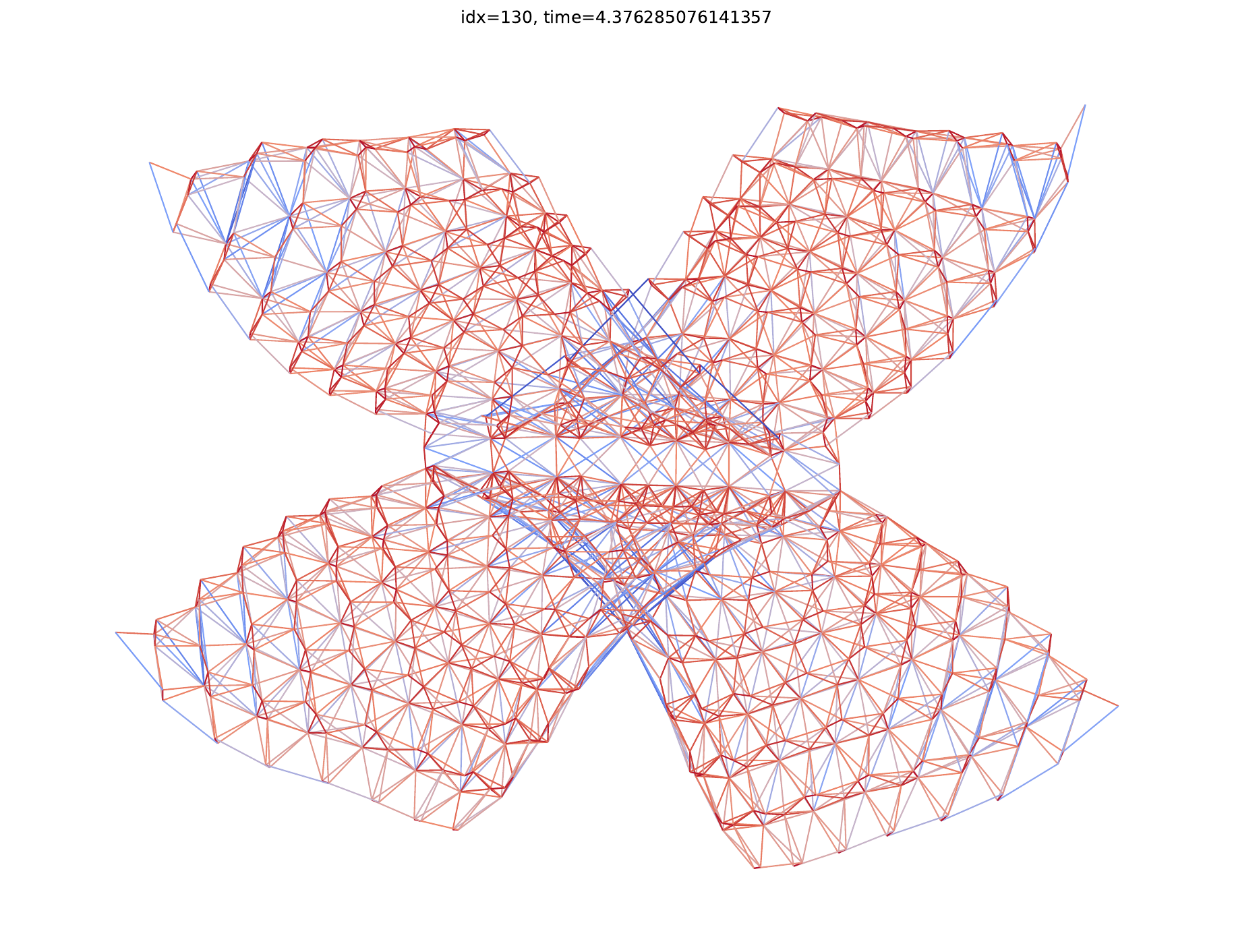} \\

&
t = 0.21s &
t = 20.45s &
t = 48.10s &
t = 2.52s &
t = 7200.00s &
t = 2.84s &
t = 2.60s &
t = 2.75s &
t = 2.60s &
t = 2.83s &
t = 2.68s &
t = 2.67s \\

\makecell{\bfseries can\_838\\N = 838\\M = 4586} &
\imgcell{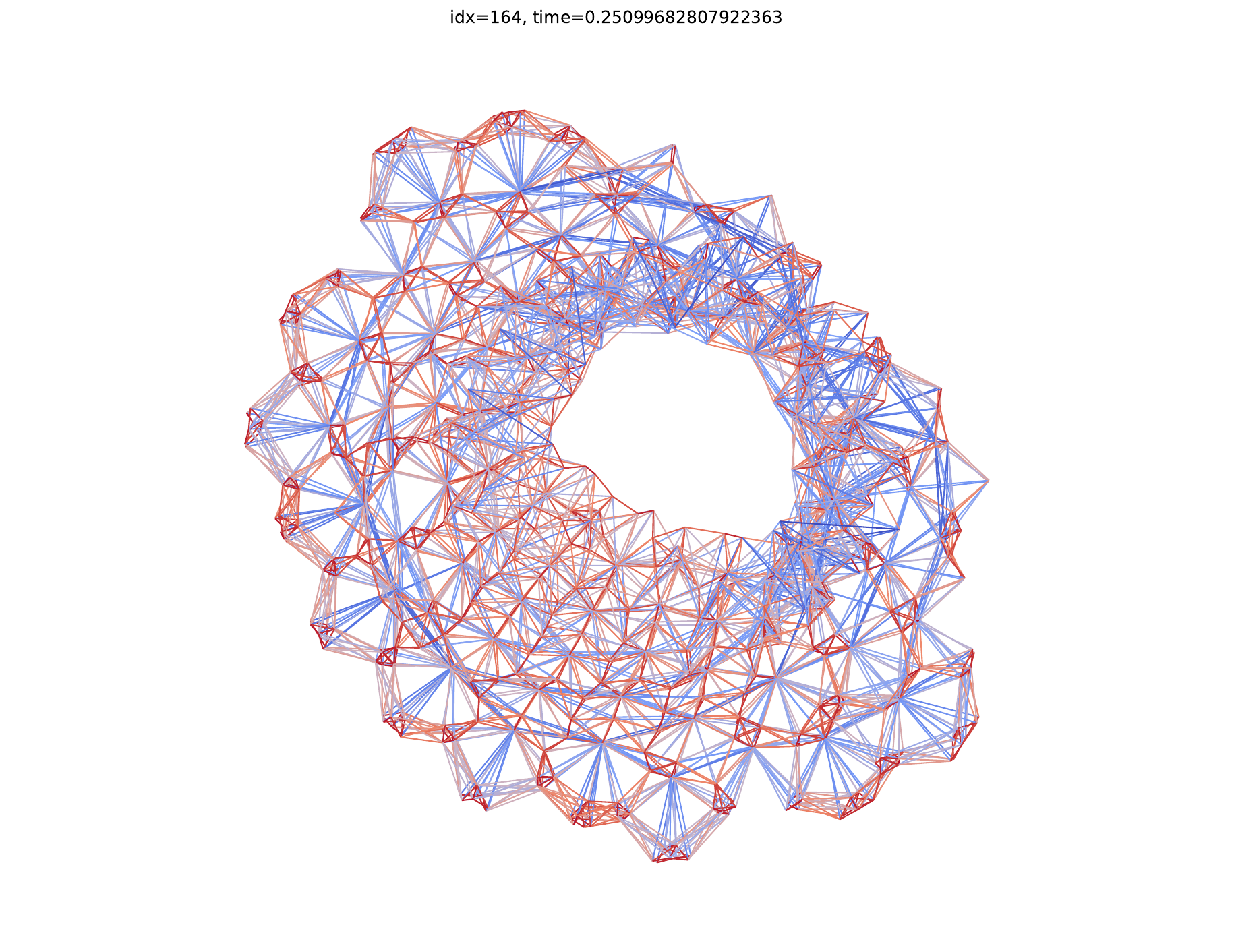} &
\imgcell{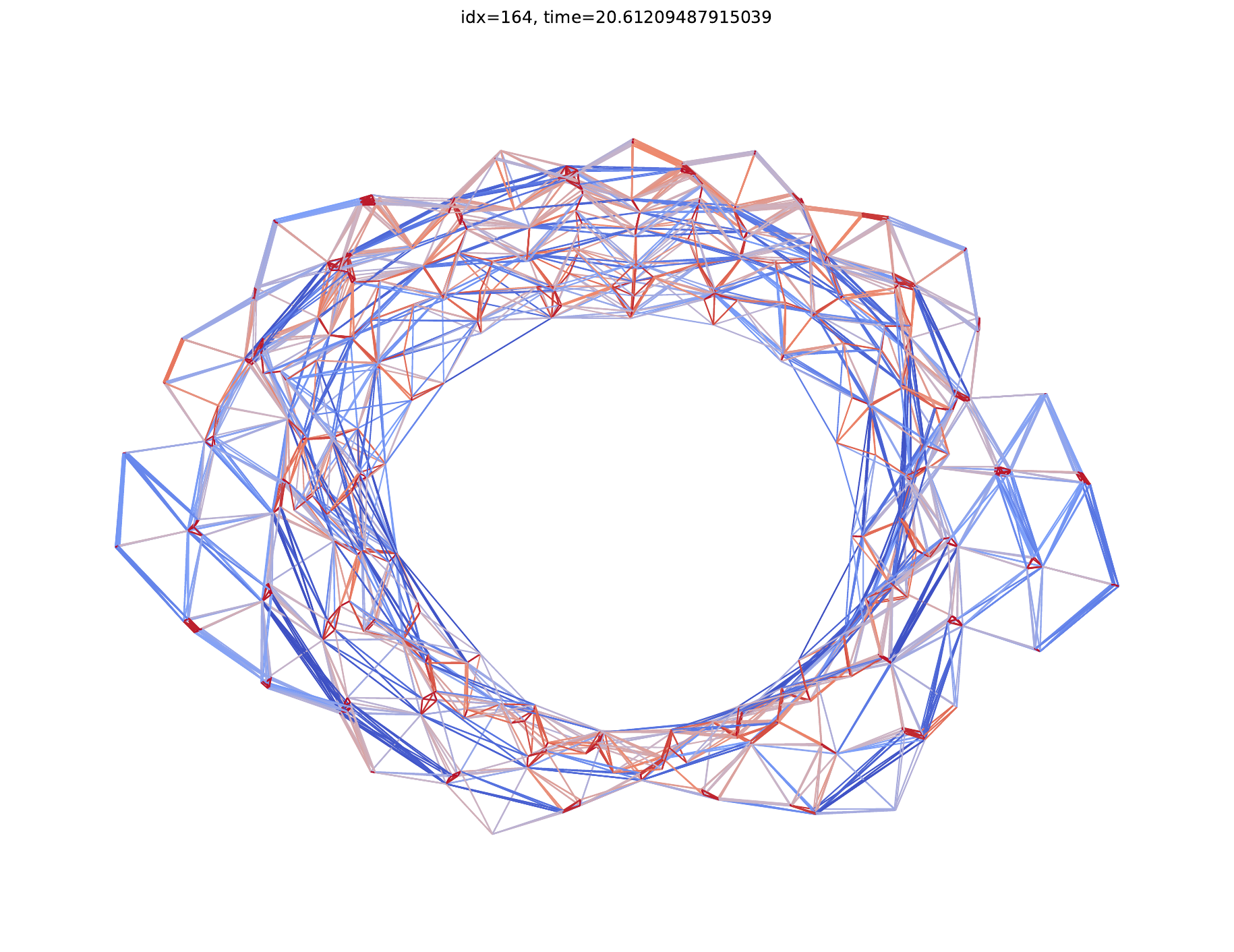} &
\imgcell{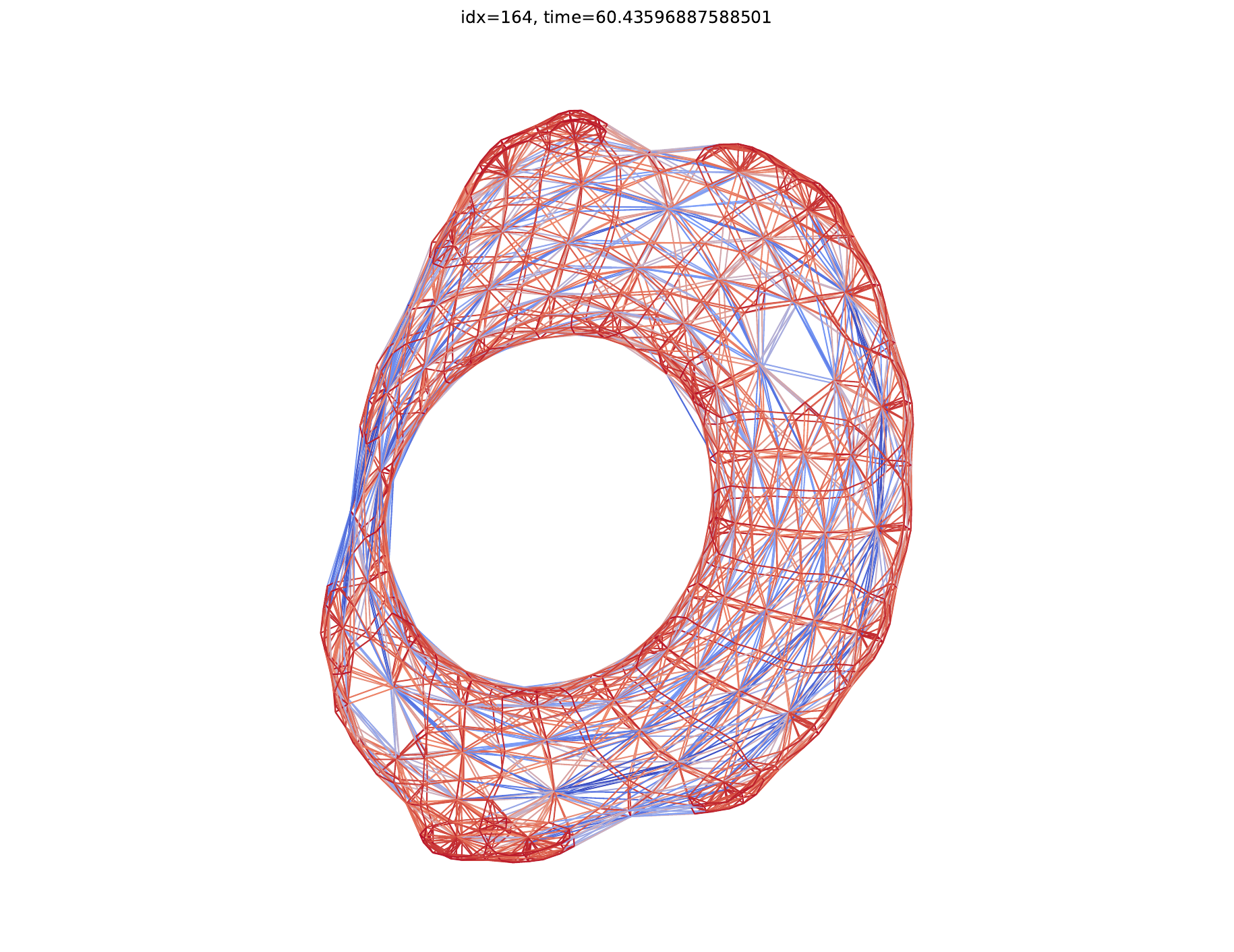} &
\imgcell{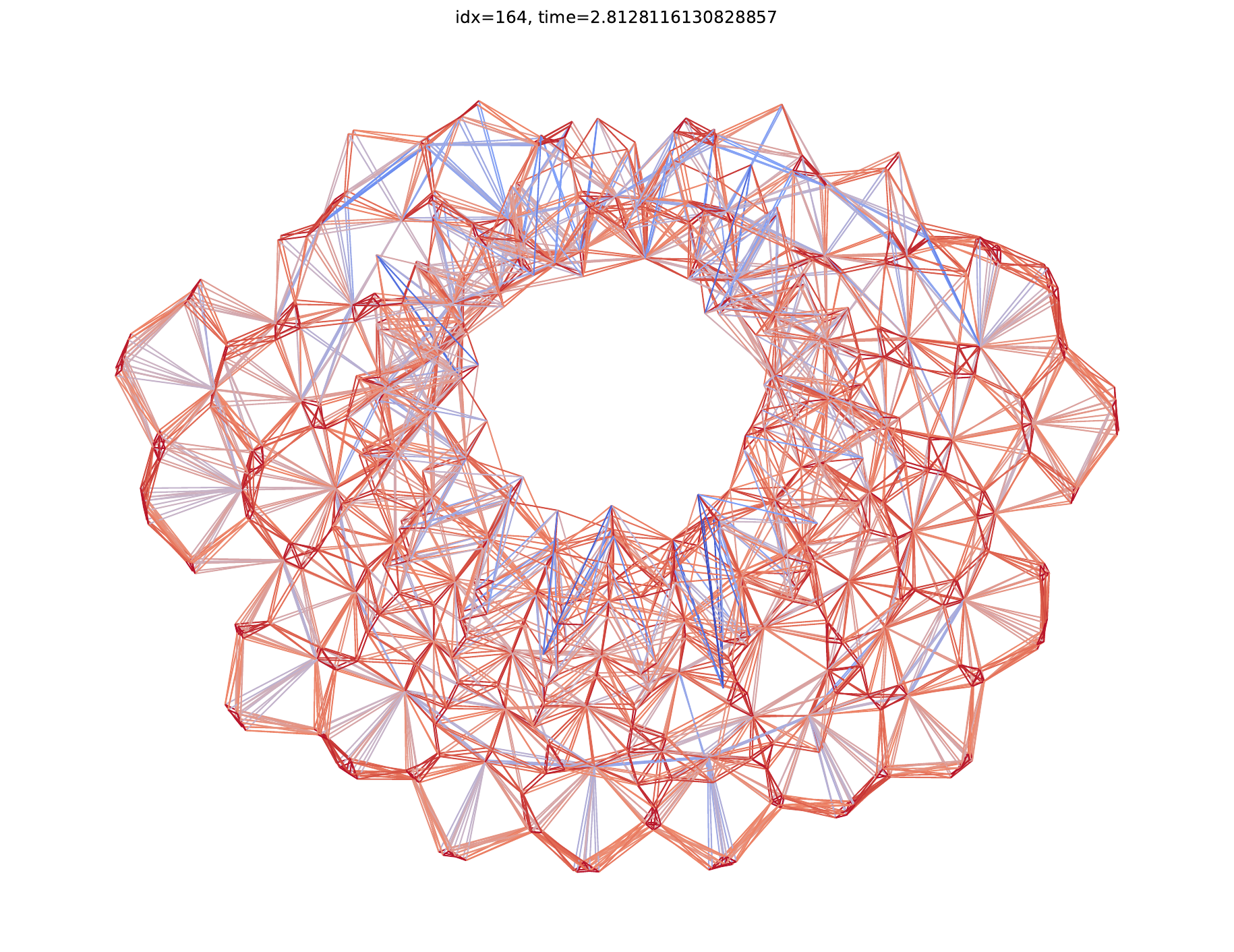} &
\imgcell{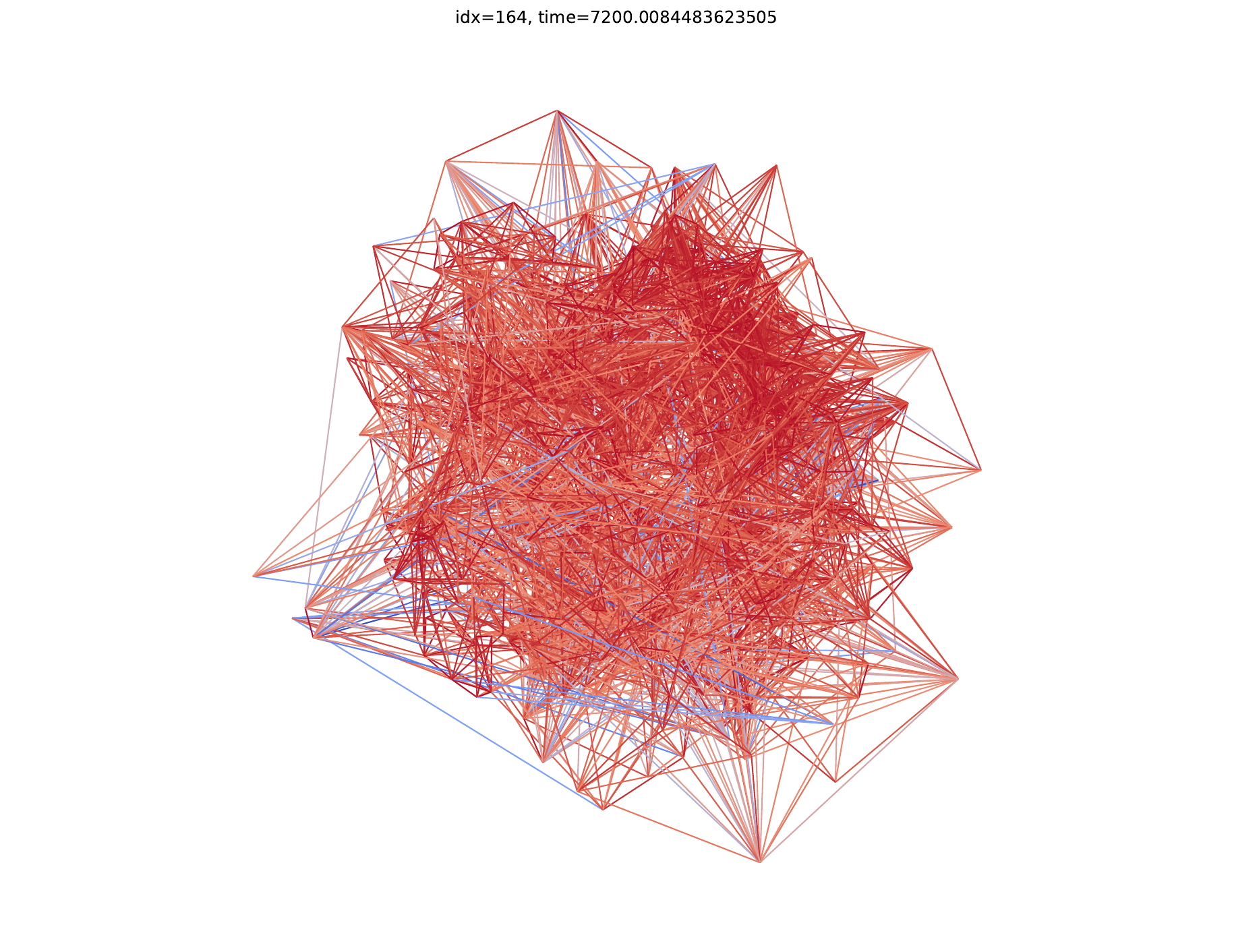} &
\imgcell{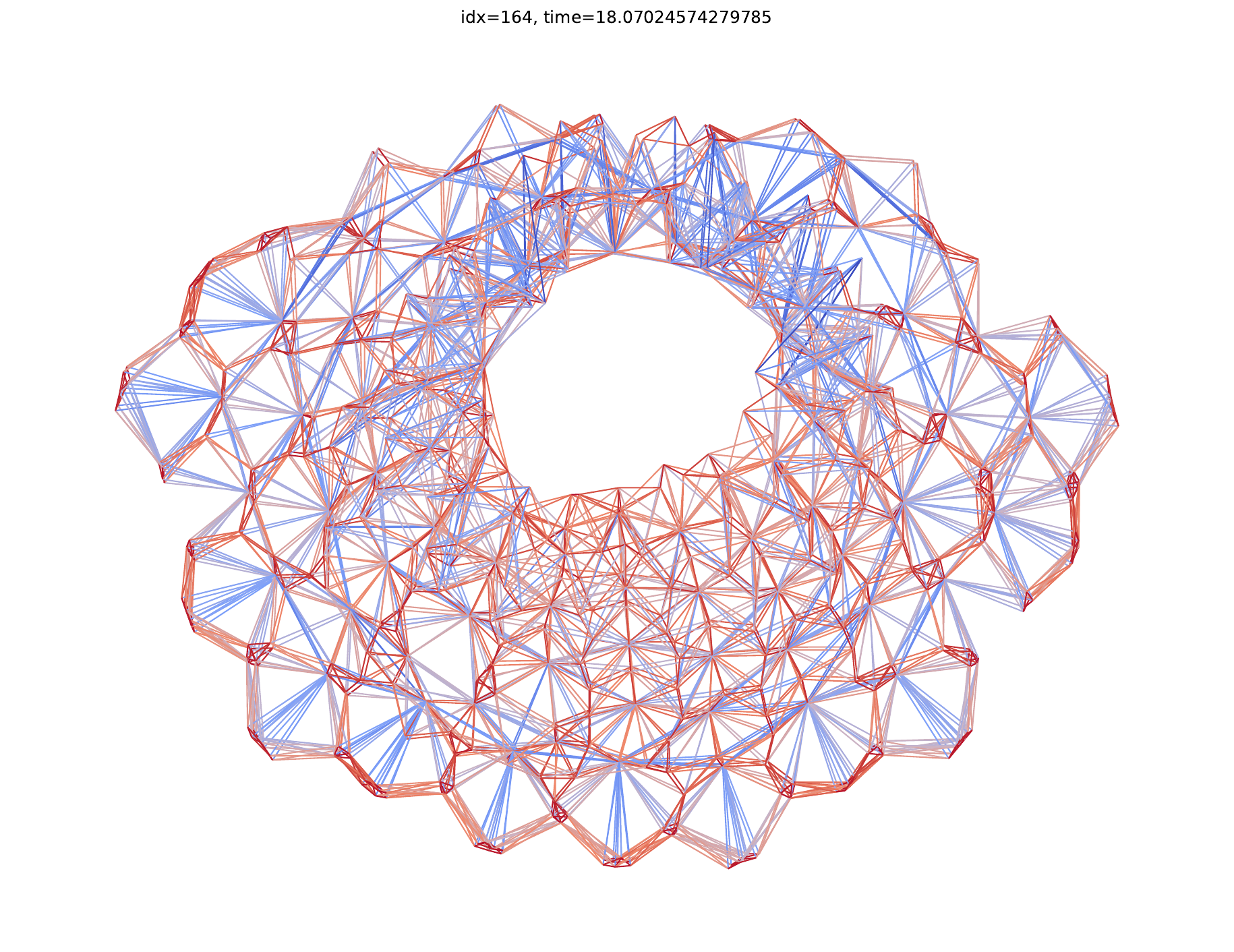} &
\imgcell{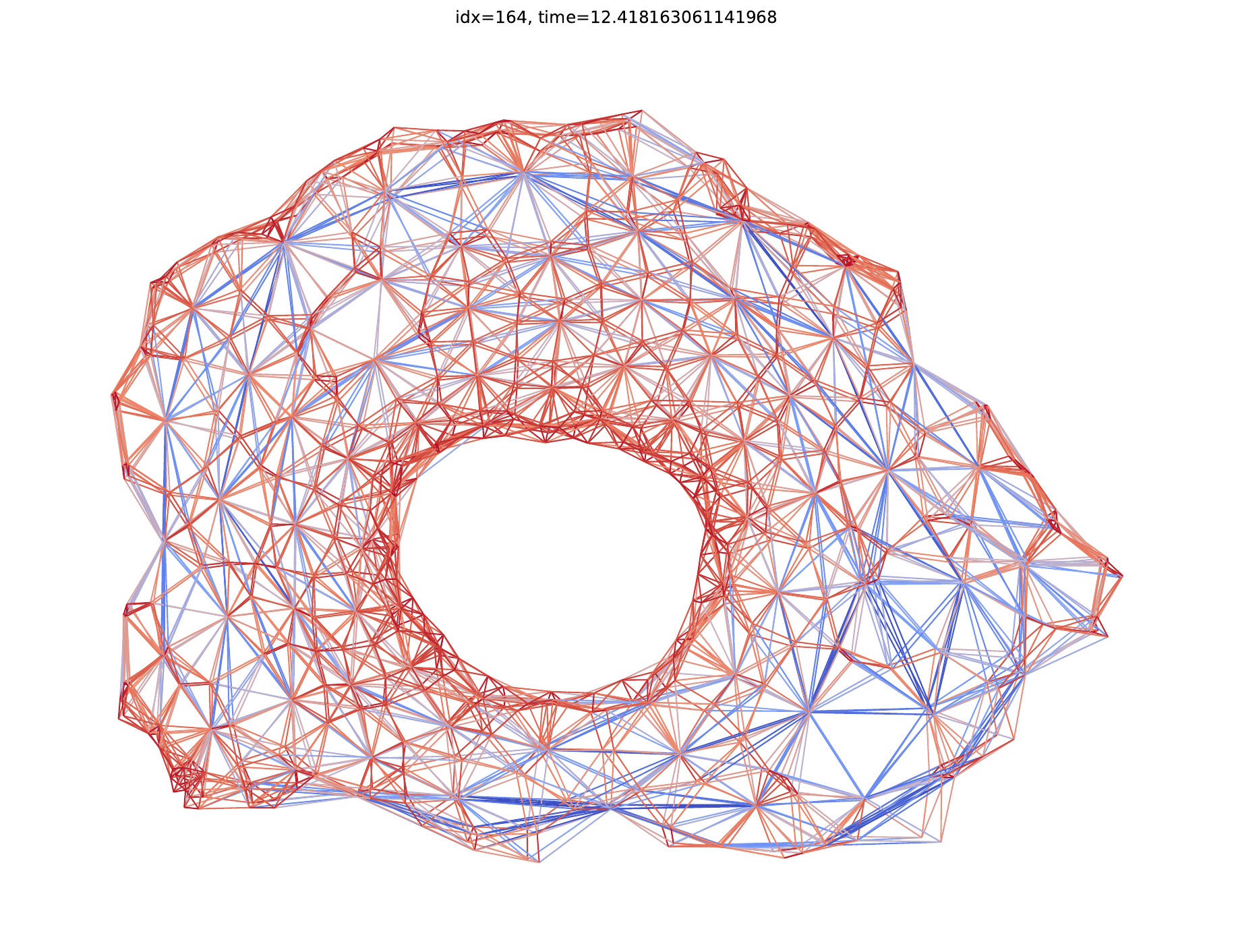} &
\imgcell{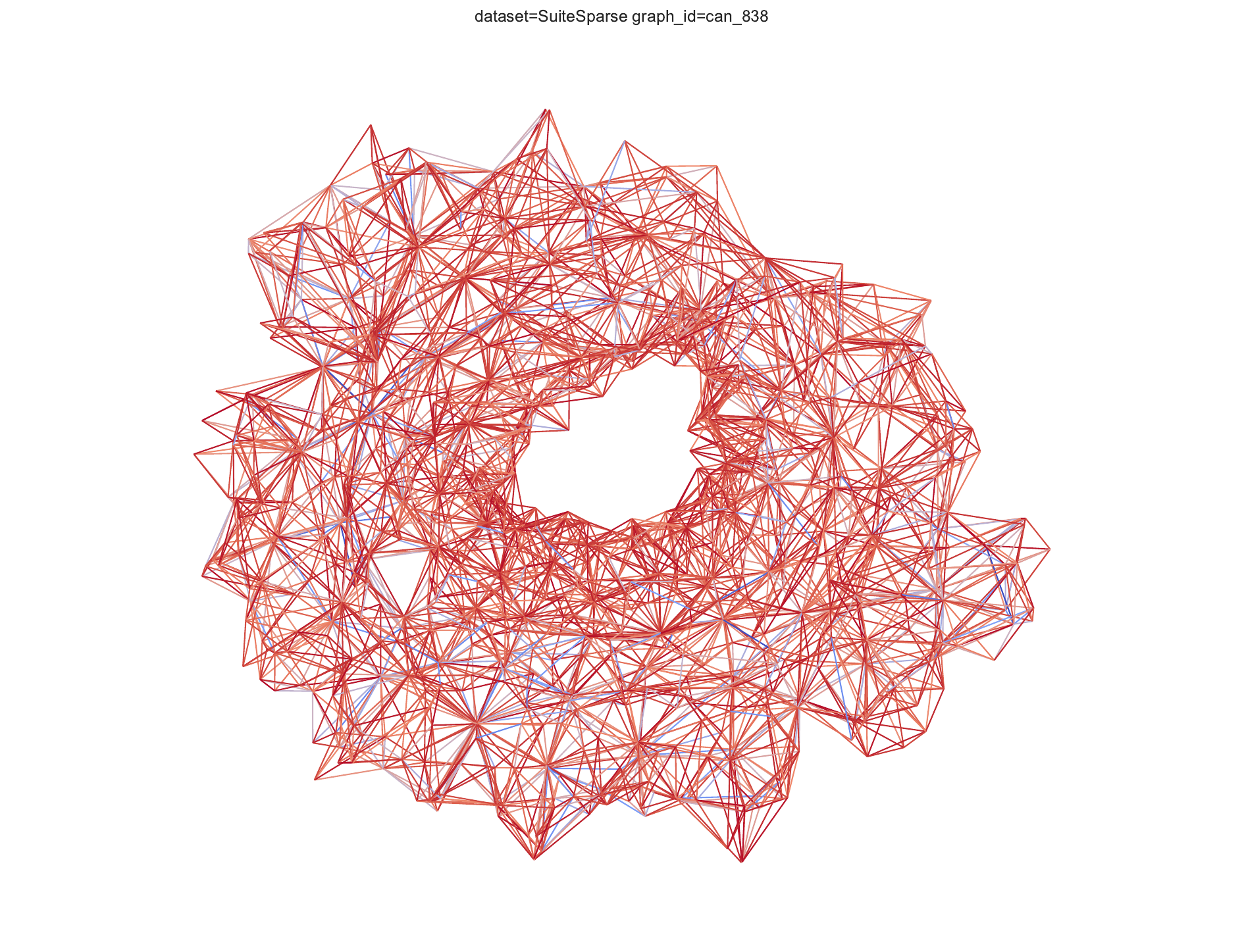} &
\imgcell{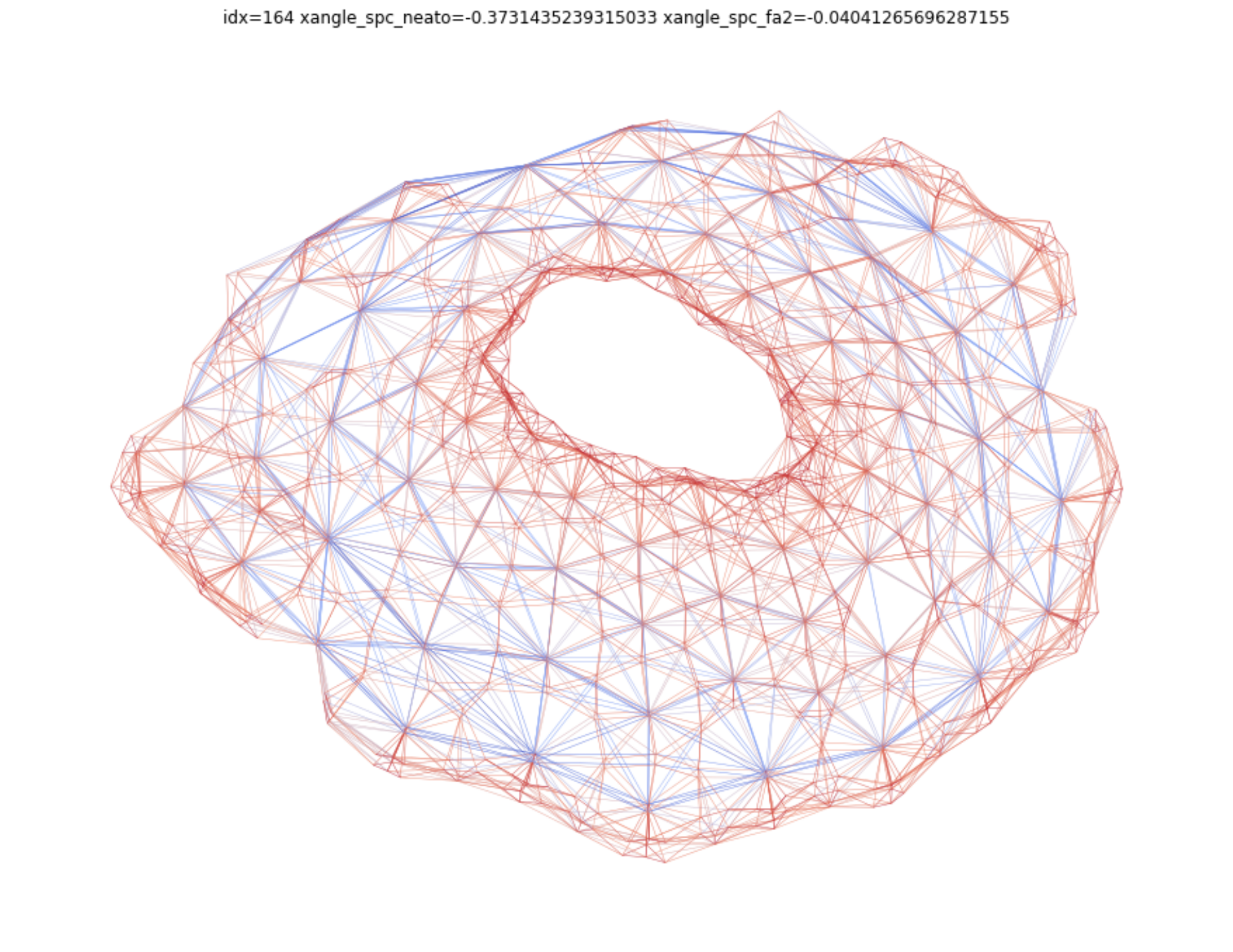} &
\imgcell{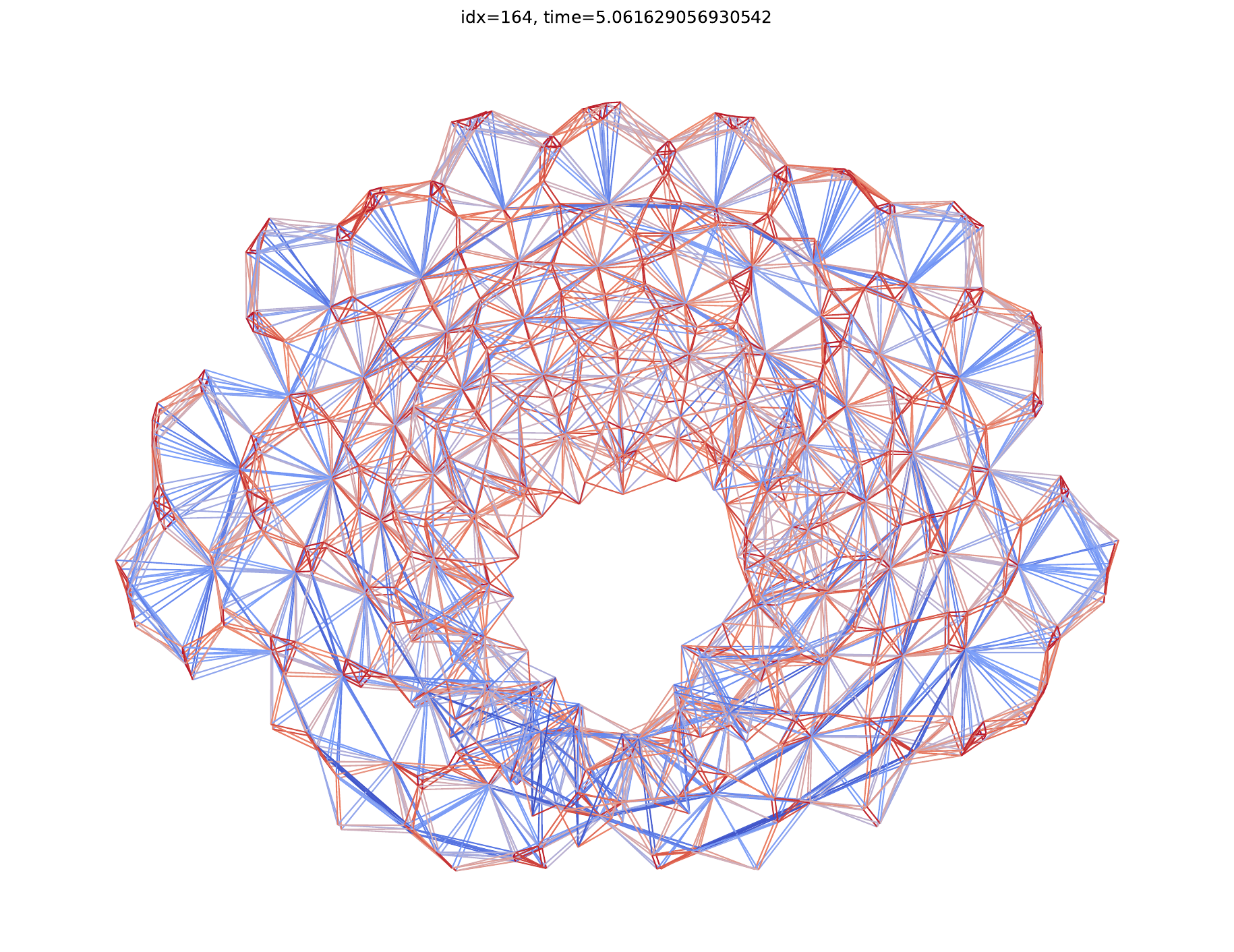} &
\imgcell{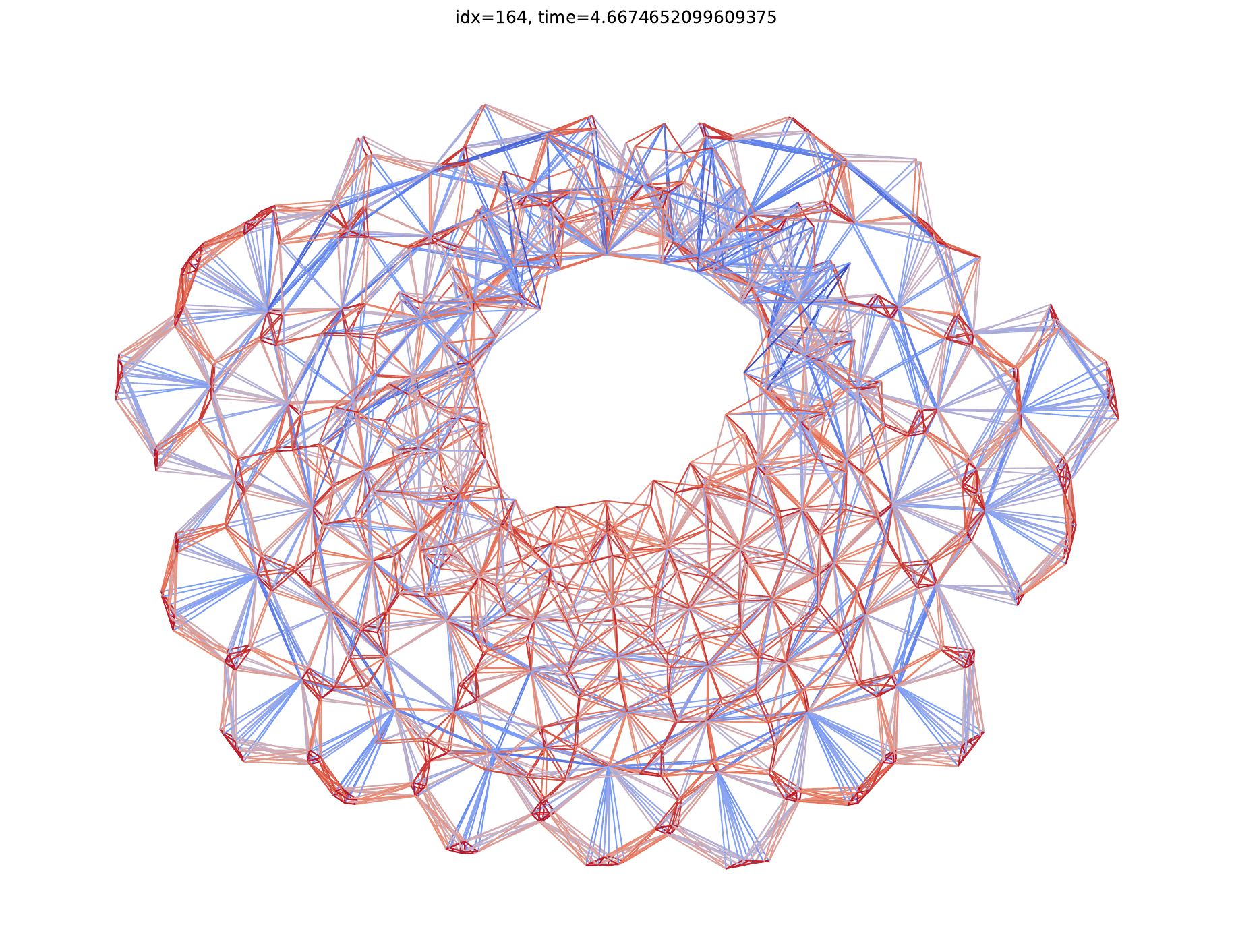} &
\imgcell{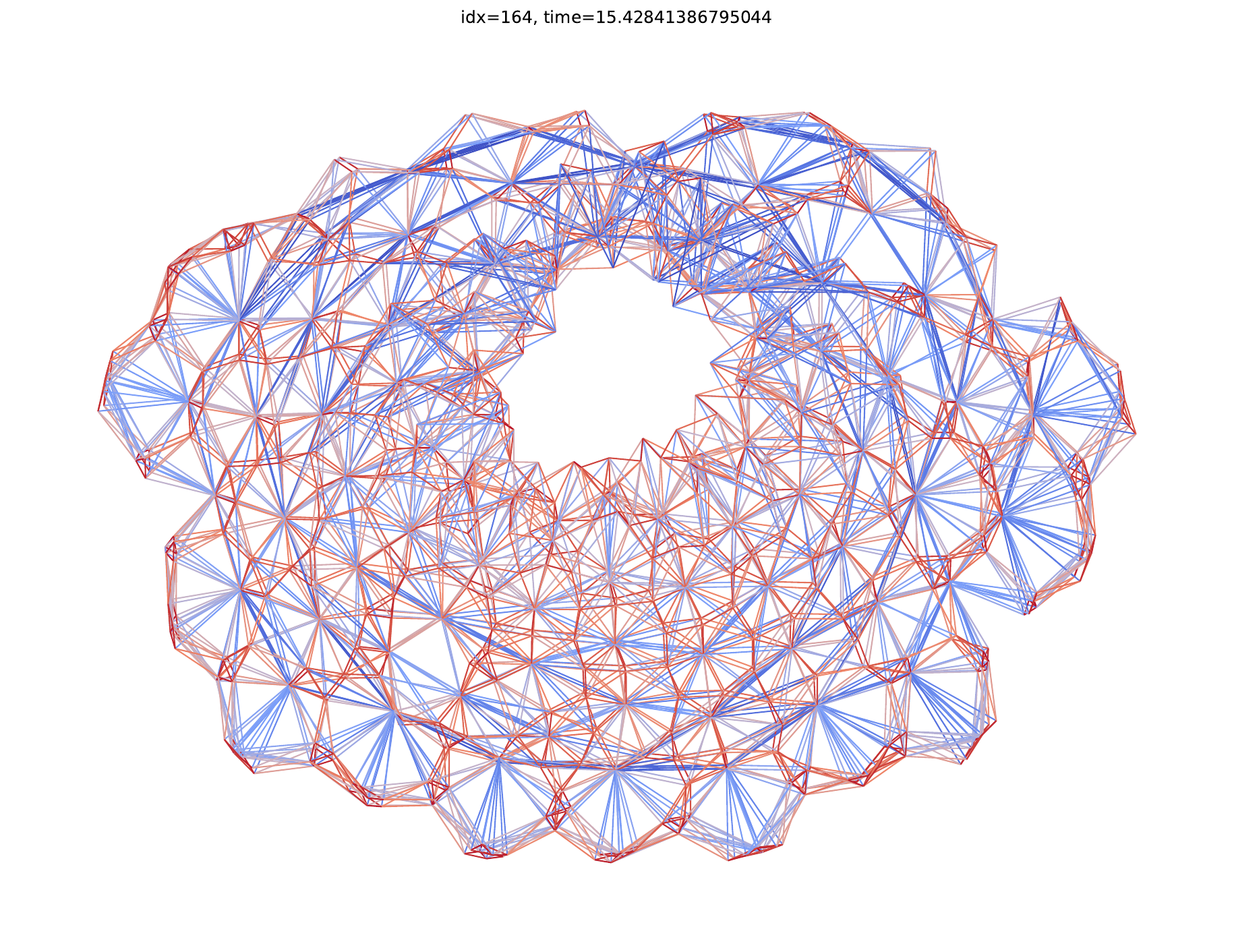} \\

&
t = 0.25s &
t = 20.61s &
t = 60.44s &
t = 2.81s &
t = 7200.00s &
t = 3.27s &
t = 3.11s &
t = 3.21s &
t = 3.01s &
t = 3.35s &
t = 2.80s &
t = 2.61s \\

\makecell{\bfseries utm1700b\\N = 1700\\M = 14626} &
\imgcell{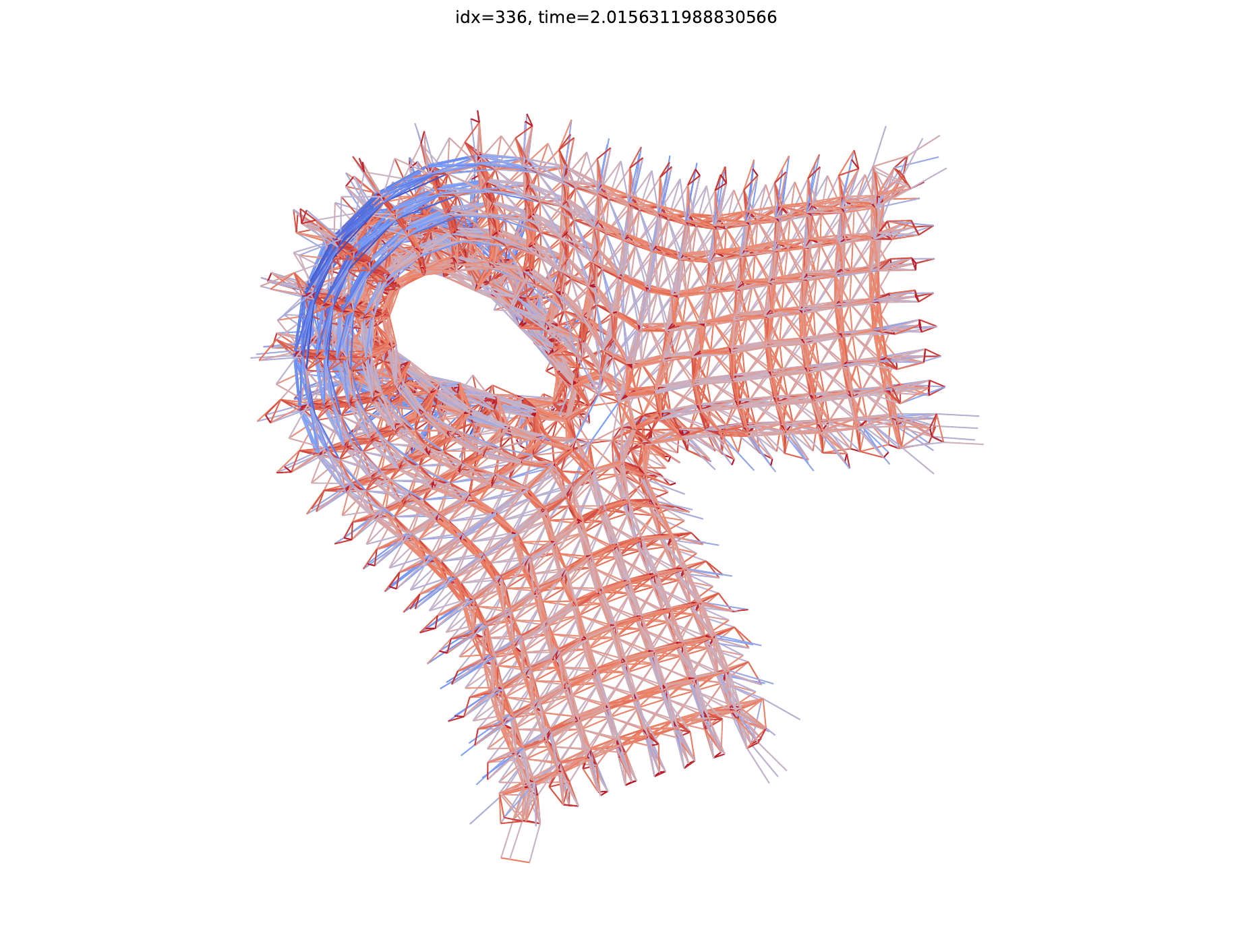} &
\imgcell{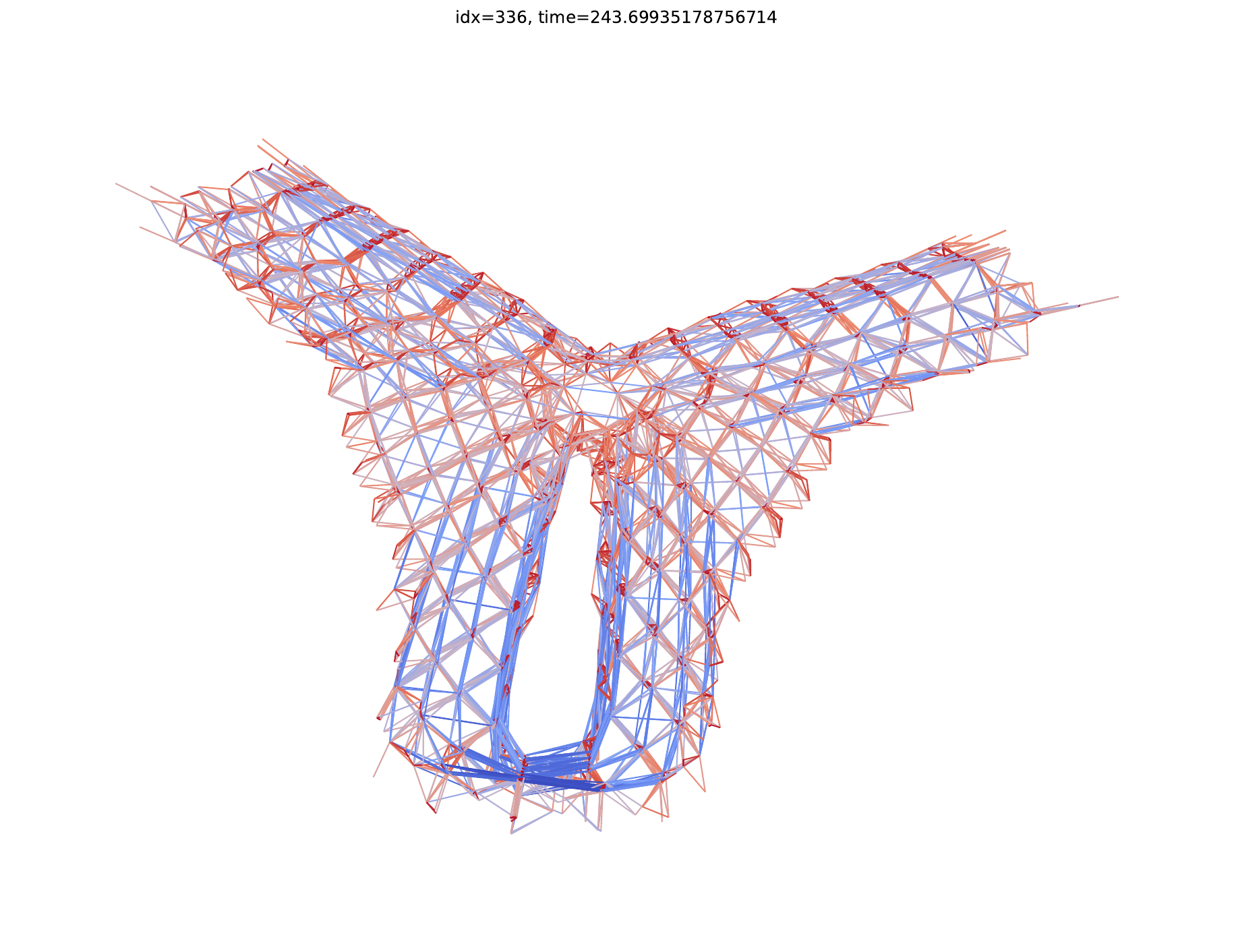} &
\imgcell{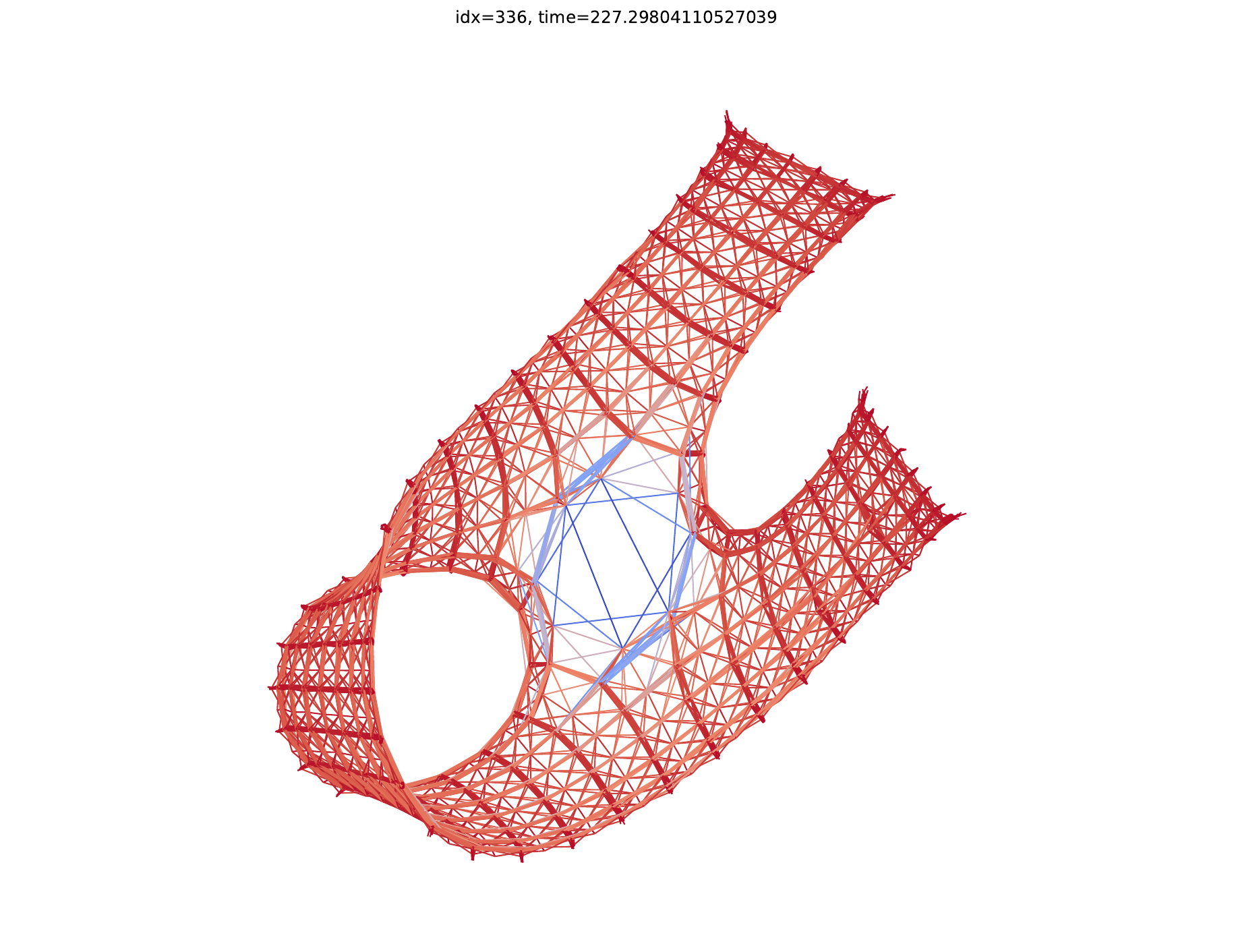} &
\imgcell{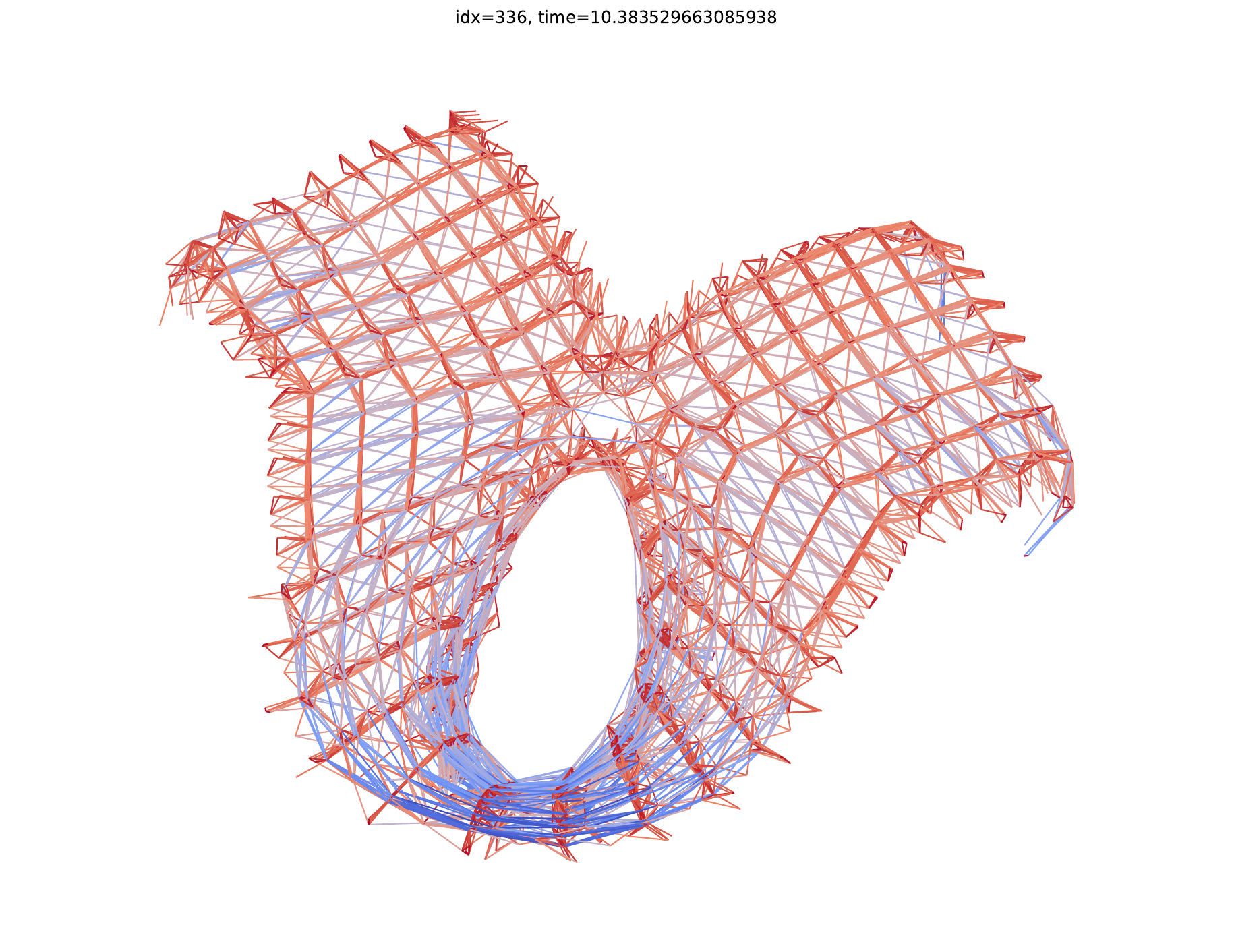} &
\imgcell{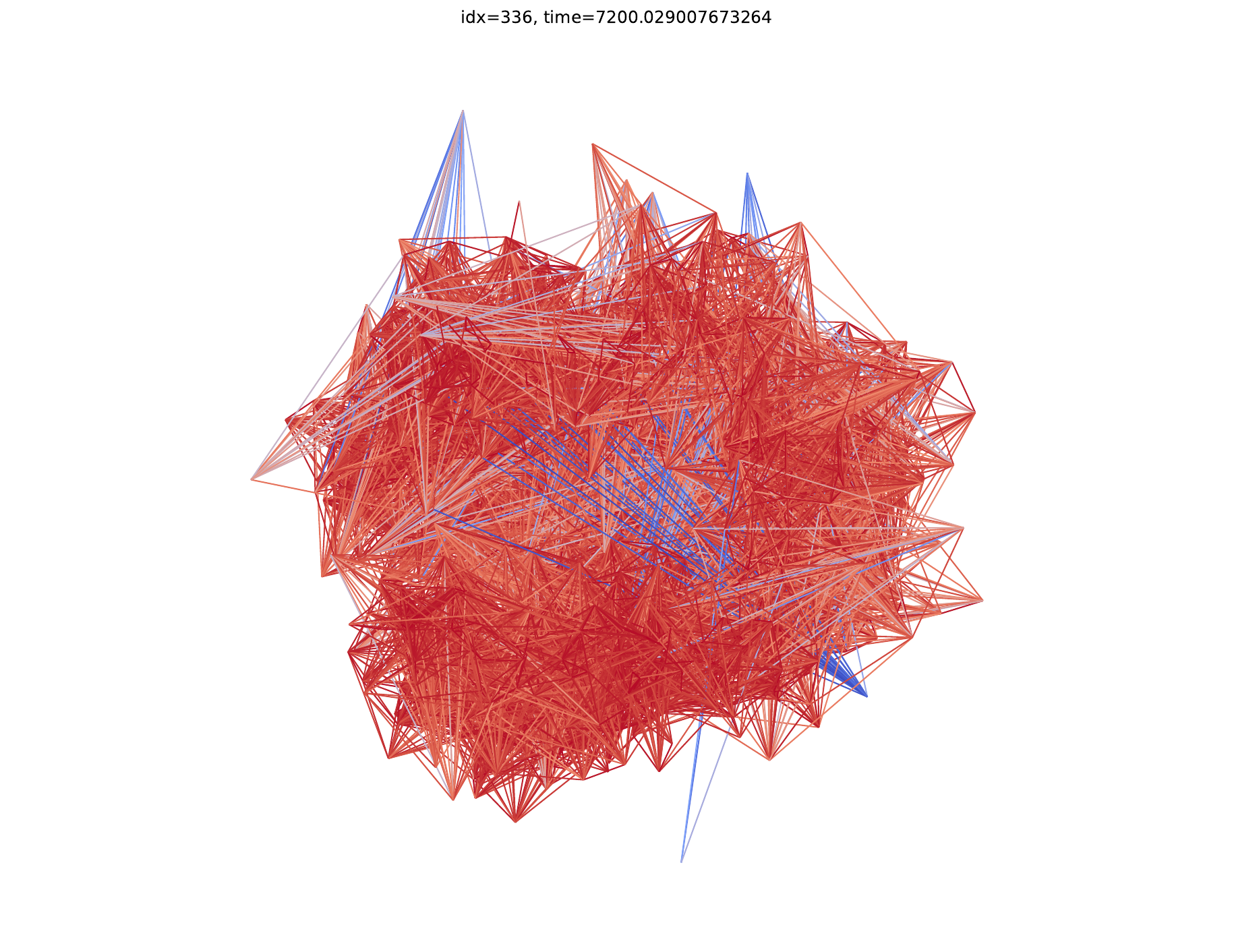} &
\imgcell{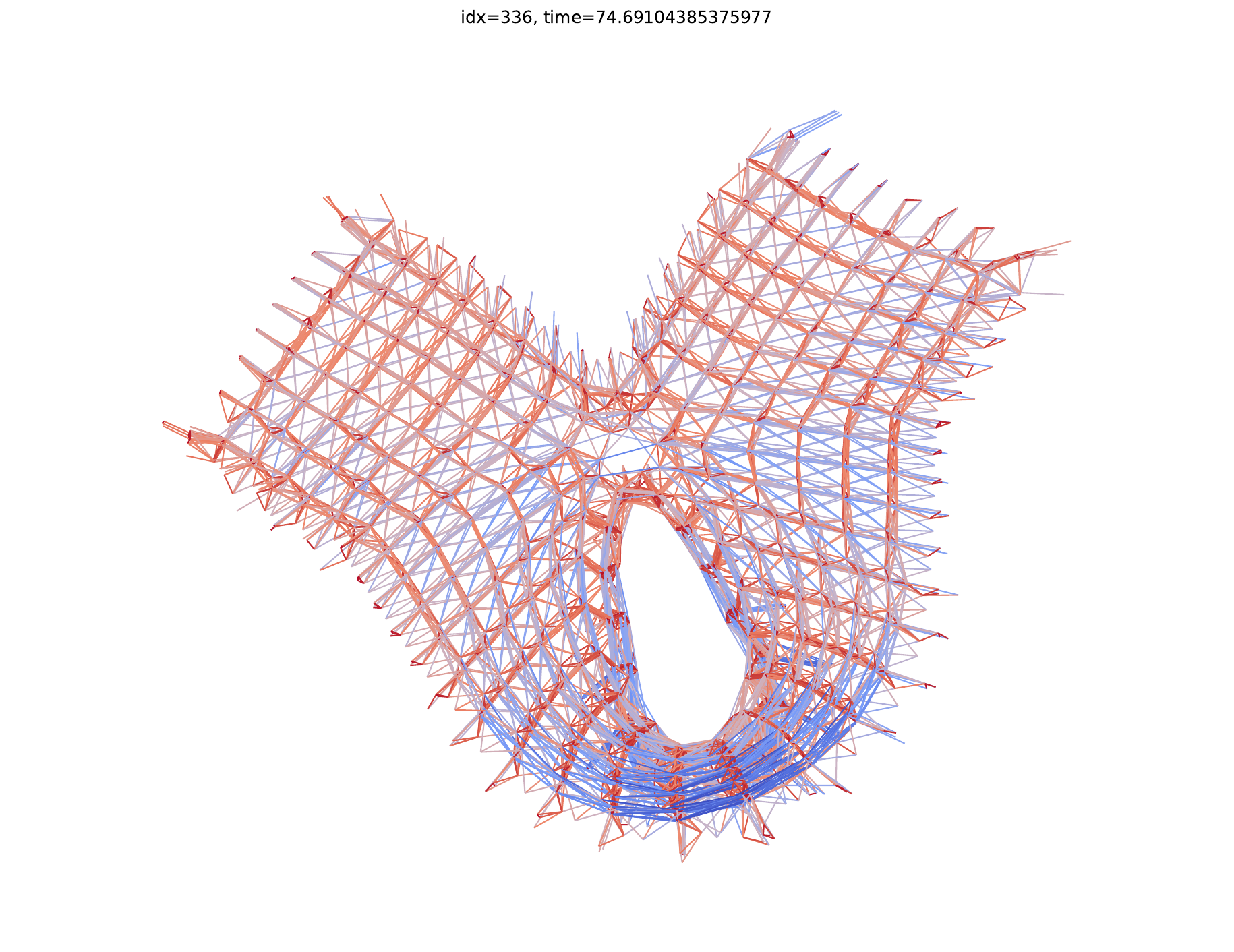} &
\imgcell{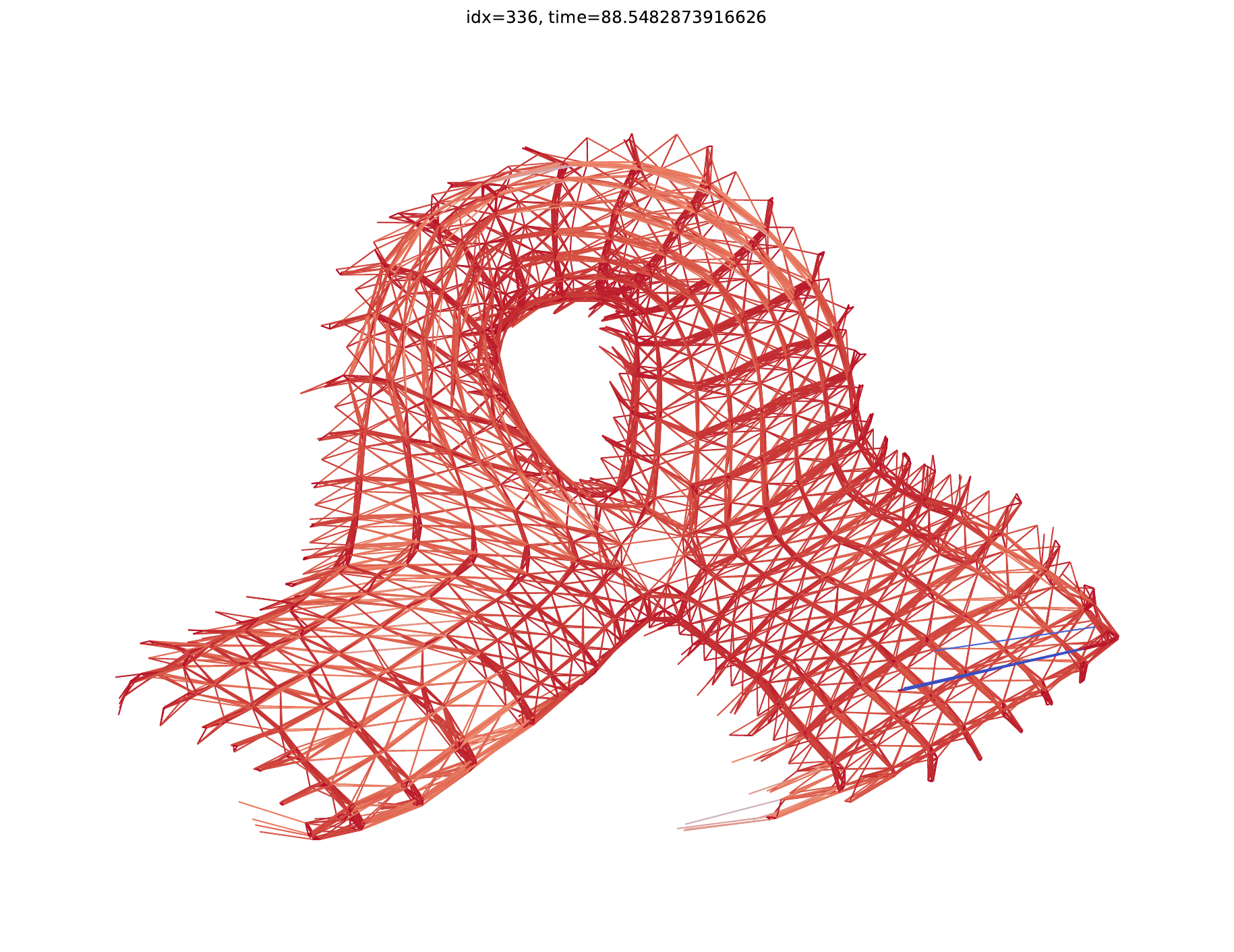} &
\imgcell{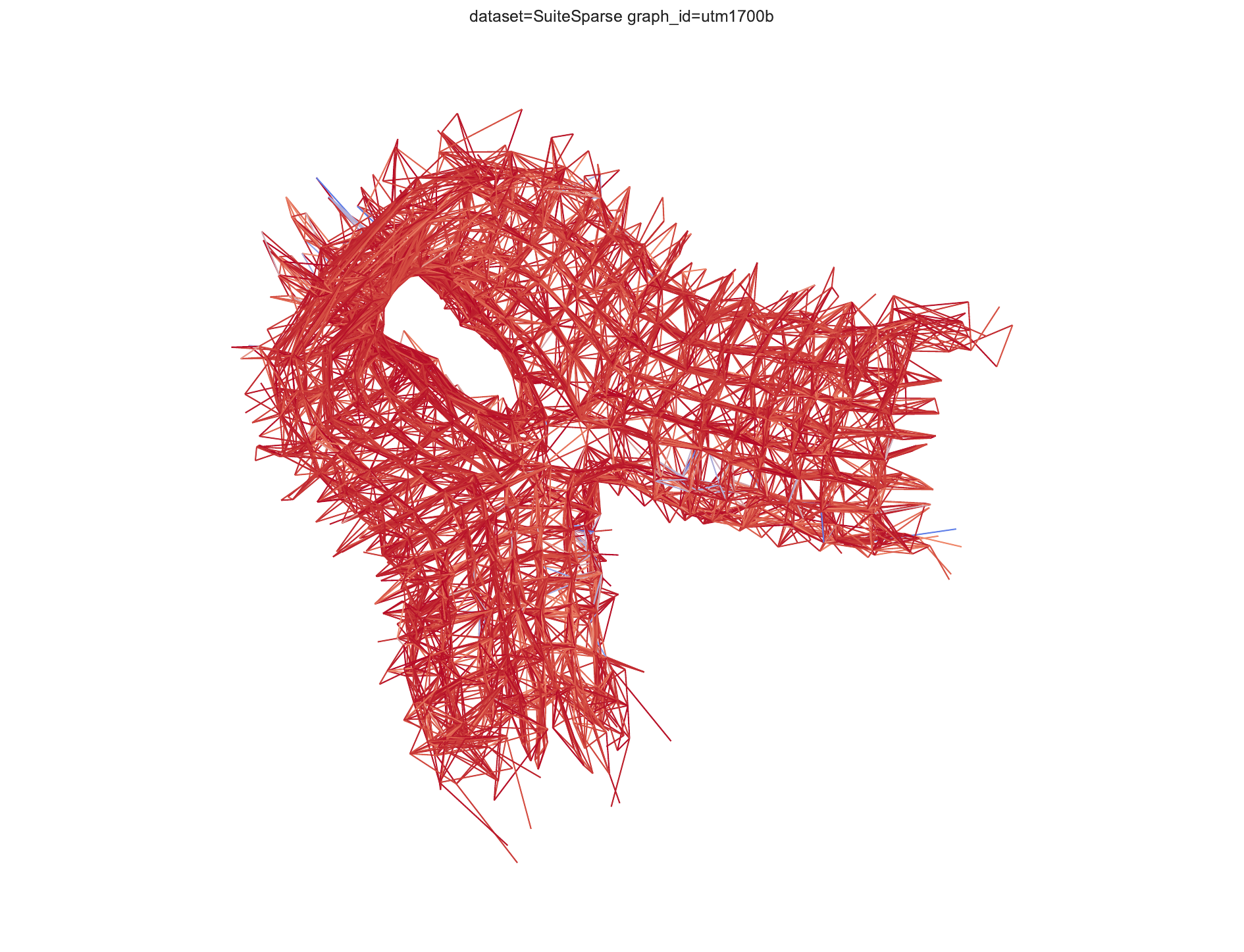} &
\imgcell{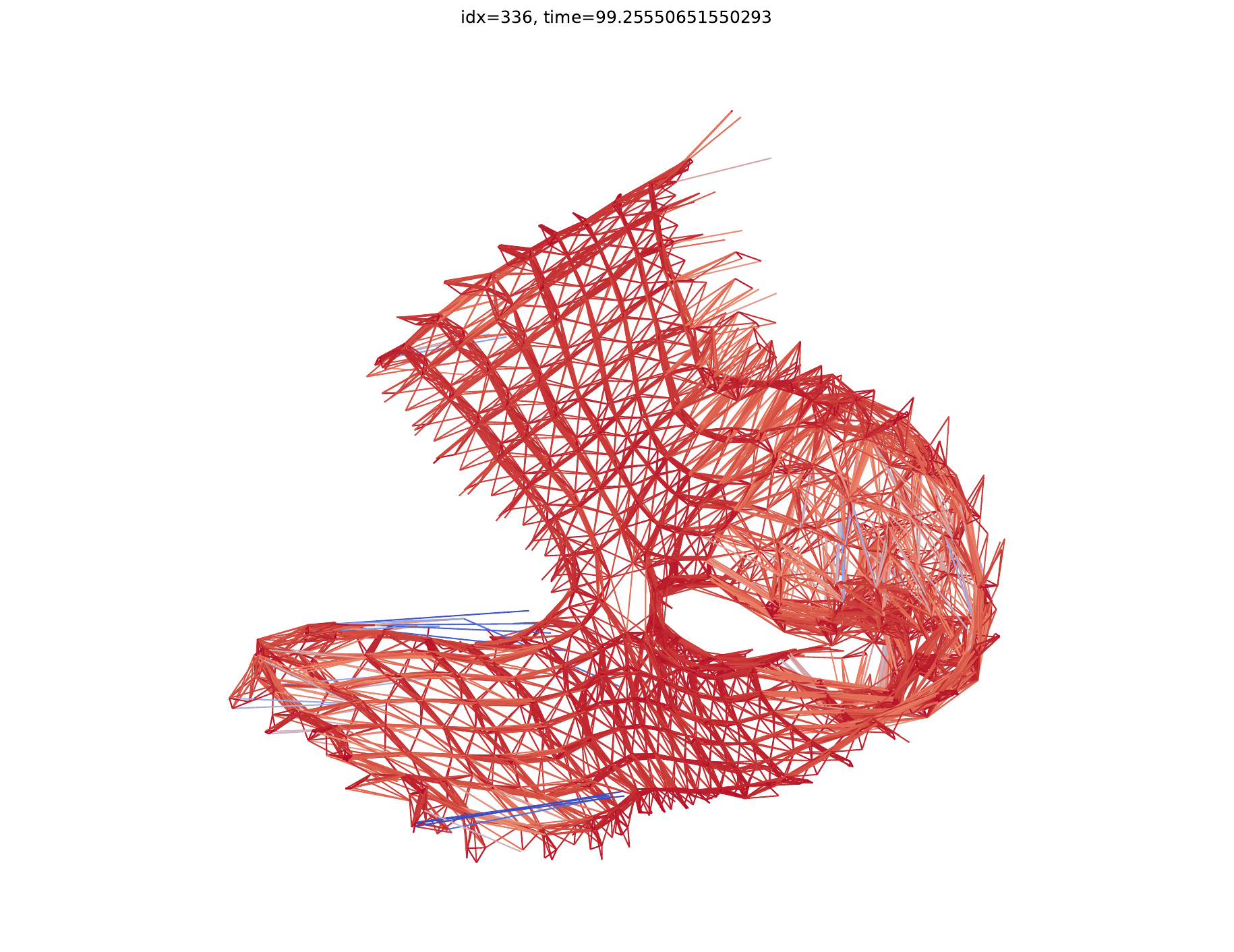} &
\imgcell{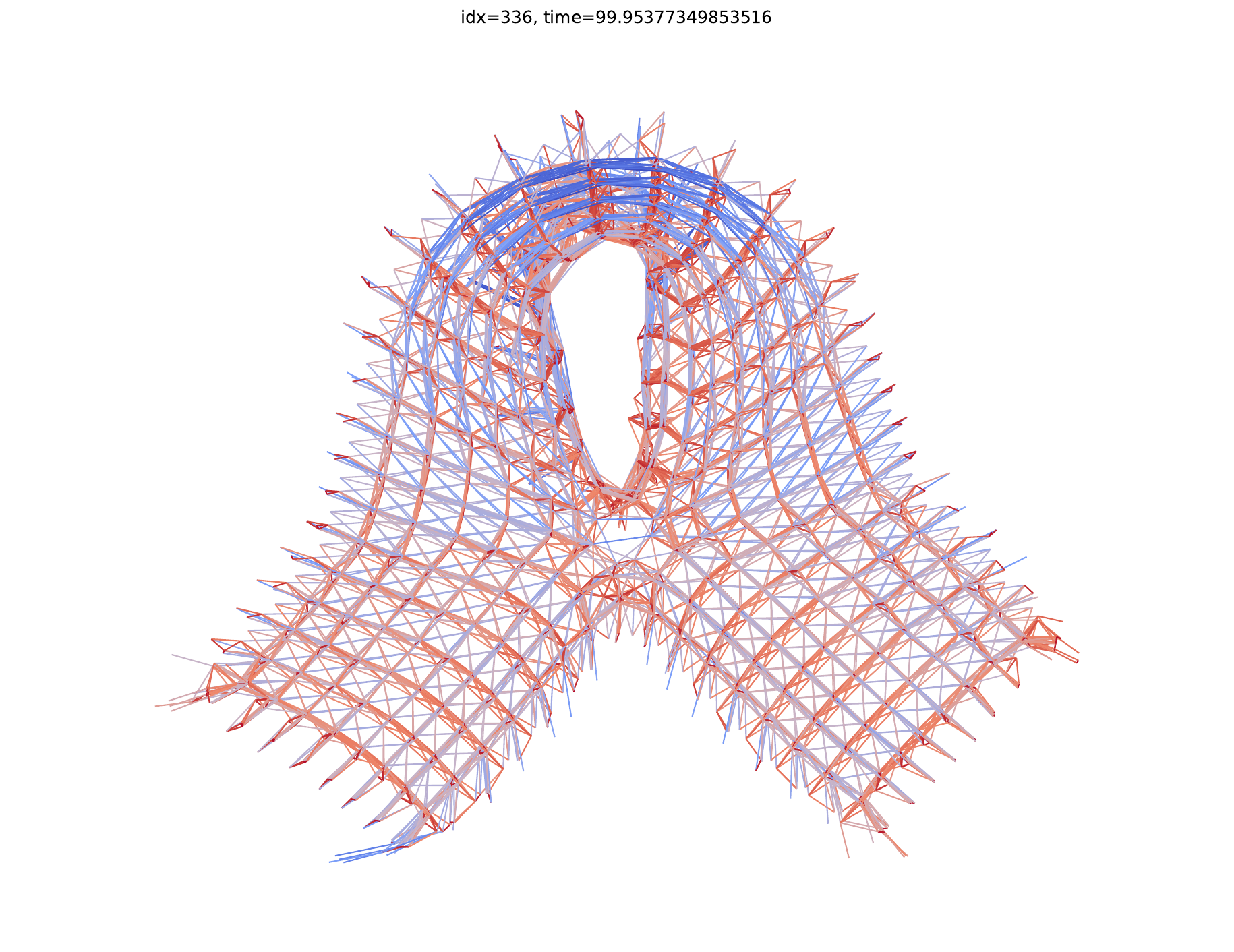} &
\imgcell{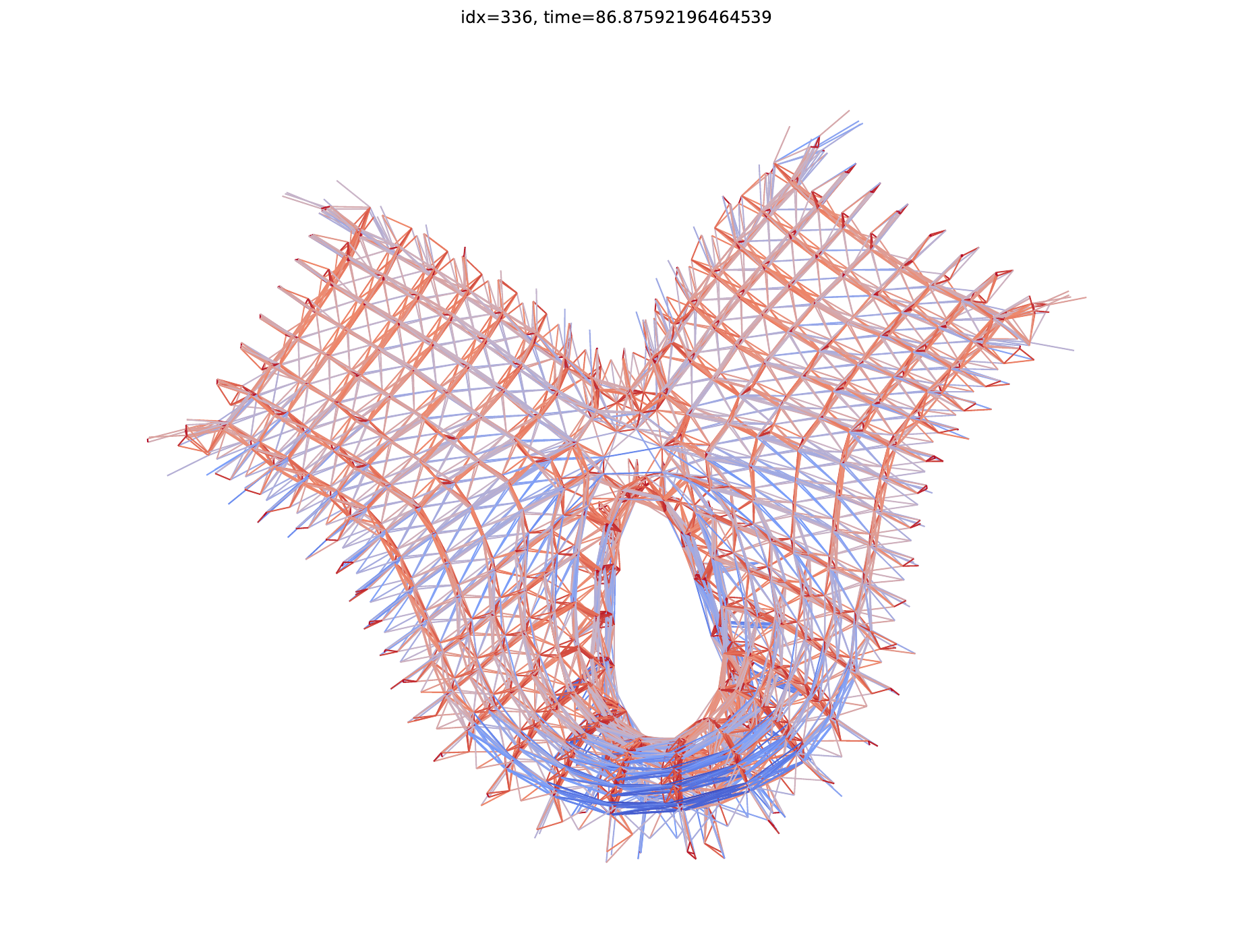} &
\imgcell{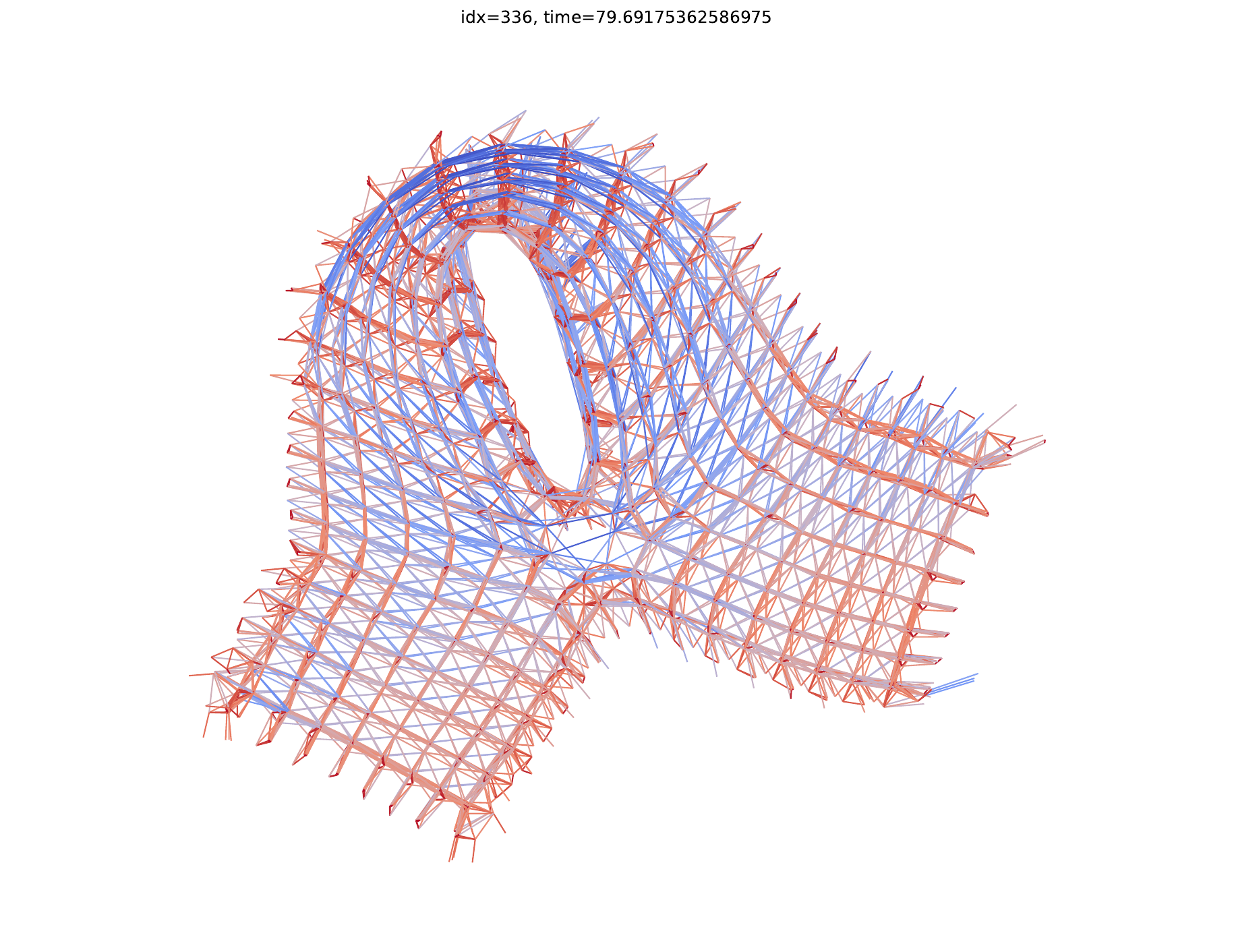} \\

&
t = 2.02s &
t = 243.70s &
t = 227.30s &
t = 10.38s &
t = 7200.00s &
t = 10.49s &
t = 8.85s &
t = 9.85s &
t = 9.96s &
t = 9.95s &
t = 10.68s &
t = 11.99s \\

\end{tabular}
\captionof{figure}[]{The qualitative evaluation of 7 \modelName\ models by comparing with 5 competitive and representative benchmarks. All the graphs presented above are unseen during the training phase of \modelName. The name of the graphs with the number of nodes $N$ and the number of edges $M$ is presented in the row header. The colors of the nodes represent their community within the graph, computed by Girvan-Newman algorithm\cite{girvan-newman}. For each layout, the computation time $t$ (without including the pre-processing time) is reported in seconds. All the computation time is computed on \minorrevise{the CPU} except the \modelName\ for Rome graphs and DeepGD for Rome graphs.}
\label{fig:vis-result}
\end{table*} 

\section{Lessons, Limitations, and Future works}

\par\revise{\textbf{Lessons.} In the process of developing this work, we have learned some important lessons regarding the effect of different hyper-parameters and the implementation of \ganName. Speaking of the effect of different hyper-parameters, we found that varying the generator size (e.g., the width and depth of hidden layers and the size of edge nets in NNConv) would not affect the model performance significantly, while the size of the discriminator is a more sensitive factor that impacts the quality of the generated layouts. A possible explanation is that the discriminator must possess the capability of correctly estimating the ratio between the data density and generator density, whereas this capability cannot be easily learned by a small network. In addition, during the development of \ganName, we observed that the timing of replacement plays a very important role in model training. In a given epoch, if we replace the good layouts as soon as we find the superior generated layouts instead of replacing them after the current epoch is finished, the generator can generate layouts of better quality. One potential reason is that the delayed replacement actually gives a wrong message to the discriminator in the current epoch so that the discriminator might not be able to always give accurate feedback to the generator.}

\par\revise{\textbf{Limitations.} Even though \modelName\ has been shown effective in generating high-quality layouts to meet diverse aesthetic goals, it still has certain limitations concerning training data collection and scalability. First of all, the quality of good layout examples has been shown very crucial to \modelName\ performance. However, collecting high-quality training data for different aesthetic goals can sometimes be time-consuming and challenging. In the future, we hope to resolve this issue by adopting the warm restart technique\cite{warm-start} when the generator falls into the local minima. Hopefully, even without the good layout examples, the generator can consistently generate layouts better than the current good layout examples. In this way, we can rely on \ganName\ to continuously enhance the quality of the layout examples, moving closer to the optimization goal without interruption. Regarding the scalability, \modelName\ is only trained on graphs with less than 100 nodes in our current experiment. Therefore, the model performance on drawing large graphs is not always guaranteed to be good. However, due to the constraint on time and memory, training \modelName\ on graphs with more than thousands of nodes is still a challenging task. In the future, we would like to explore options for making our method more scalable.}

\par\revise{\textbf{Future Work.} Lastly, our experimental studies also indicate a promising direction for future research regarding learning abstract aesthetic goals. To be specific, we found that \modelName\ without \ganName\ can learn the implicit layout preference from a good layout collection without explicitly knowing the inherent layout preference in this collection. This finding shows a promising future in that this GAN-based approach might be able to learn to draw human-preferred layouts if the layouts in the good layout collection can be selected according to human preference.}

\section{Conclusion}
We propose \modelName, a novel Generative Adversarial Network (GAN) based graph drawing framework for diverse aesthetic goals. \red{\modelName\ can effectively generate high-quality layouts to optimize different quantitative criteria or a combination of quantitative criteria. Compared with the existing layout methods that focus on optimizing multiple aesthetics, \modelName\ addresses the unique challenge of optimizing non-differentiable criteria \minorrevise{without the need to manually define} a differentiable surrogate function.} We conduct experiments to evaluate the effectiveness of \modelName\ quantitatively and qualitatively against several widely used layout methods. The quantitative evaluation demonstrates that \modelName\ can consistently generate layouts that are of equal or better quality compared to the benchmarks. The qualitative evaluation shows that the layouts generated by \modelName\ are visually pleasing and informative. Lastly, \modelName\ can generate layouts at a relatively low computational cost compared to other methods.

\bibliographystyle{IEEEtran}


\bibliography{reference}
%









\vspace{-50pt}
\begin{IEEEbiography}[{\includegraphics[width=1in,height=1in,clip,keepaspectratio]{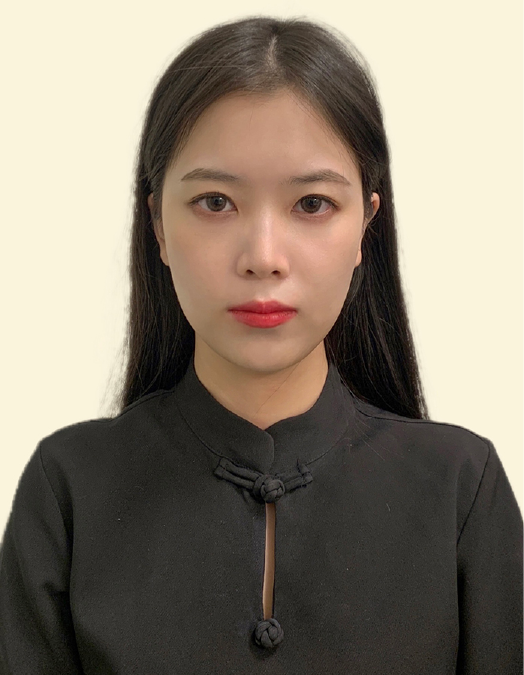}}]{Xiaoqi Wang}is a Ph.D. student in computer science at the Ohio State University. She received a Bachelor of Science in data analytics from the Ohio State University and a Master of Science in data science from Columbia University. Her research interests include graph drawing and AI explainability. Contact her at wang.5502@osu.edu.
\end{IEEEbiography}
\vspace{-50pt}

\begin{IEEEbiography}[{\includegraphics[width=1in,height=1in,clip,keepaspectratio]{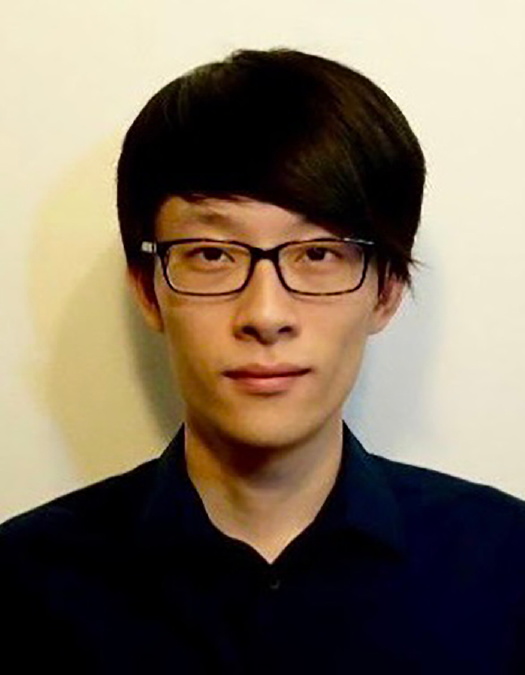}}]{Kevin Yen}is a Research Engineer at Yahoo Research. He works on various system design and development that involves applying Natural Language Processing, Computer Vision, and machine learning to products and services. Contact him at kevinyen@verizonmedia.com.
\end{IEEEbiography}
\vspace{-50pt}

\begin{IEEEbiography}[{\includegraphics[width=1in,height=1in,clip,keepaspectratio]{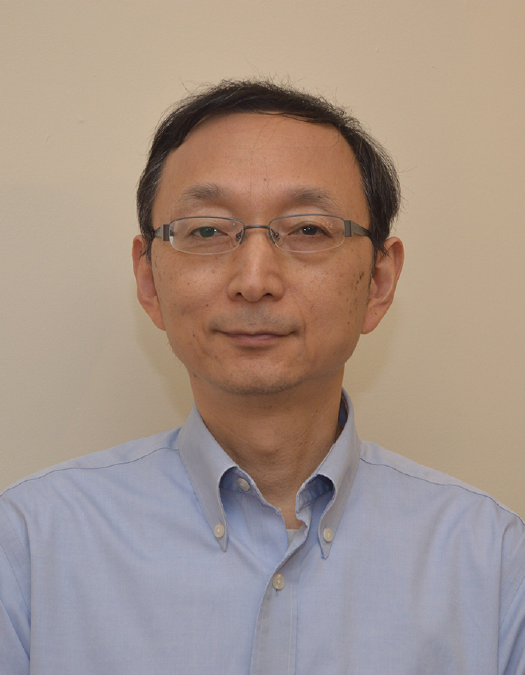}}]{Yifan Hu}is a Senior Director of Research at Yahoo Research. Prior to joining Yahoo, he worked at AT\&T Labs, Wolfram Research, and Daresbury Laboratory. He received his B.S. and M.S. in applied mathematics from Shanghai Jiao-Tong \minorrevise{University and Ph.D.} in optimization from Loughborough University. His research interests include data mining, \minorrevise{machine learning, and visualization.} He is a co-author of a number of best papers, including the 2017 ICDM 10-year highest impact award paper on recommender systems. He is the author of a number of functions in Mathematica and contributes to the \minorrevise{open-source} software Graphviz. Contact him at yifanh@gmail.com.
\end{IEEEbiography}
\vspace{-40pt}

\begin{IEEEbiography}[{\includegraphics[width=1in,height=1in,clip,keepaspectratio]{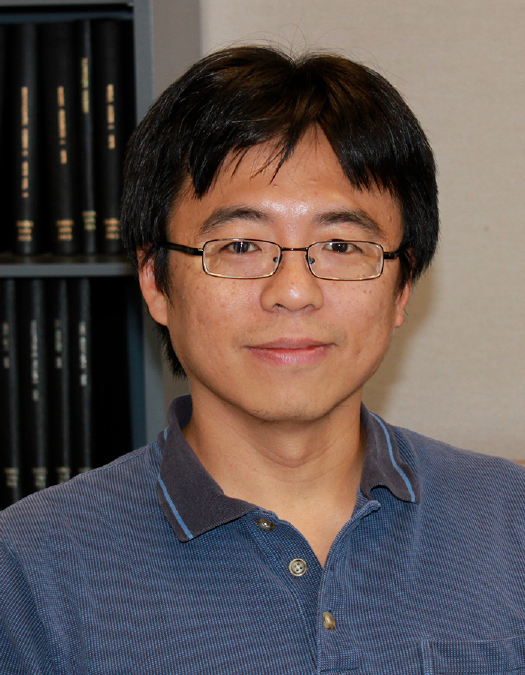}}]{Han-Wei Shen}is a Full Professor at The Ohio State University. He is an inductee of \minorrevise{the IEEE Visualization Academy.} His primary research interests are scientific visualization and computer graphics. Professor Shen is a winner of \minorrevise{the National Science} Foundation's CAREER award \minorrevise{and the US Department of Energy's Early Career Principal Investigator Award}. He received his BS degree from \minorrevise{the Department of Computer Science} and Information Engineering at National Taiwan University in 1988, the MS degree in computer science from the State University of New York at Stony Brook in 1992, and the \minorrevise{Ph.D.} degree in computer science from the University of Utah in 1998. From 1996 to 1999, he was a research scientist at NASA Ames Research Center \minorrevise{in Mountain View}, California. Contact him at shen.94@osu.edu.
\end{IEEEbiography}
\vspace{-40pt}

\end{document}


\newcommand{\ganName}{self-challenging GAN}

\newcommand{\modelName}{SmartGD}
\newcommand{\dis}{\Phi_\mathrm{dis}}
\newcommand{\gen}{\Phi_\mathrm{gen}}
\newcommand{\Xr}[1]{\mathbf{X}r^{(#1)}}
\newcommand{\Xf}[1]{\mathbf{X}f^{(#1)}}
\newcommand{\XrG}[2]{\mathbf{X}r^{(#1)}_{G_{#2}}}
\newcommand{\XfG}[2]{\mathbf{X}f^{(#1)}_{G_{#2}}}
%



\title{\modelName: A GAN-Based Graph Drawing Framework for Diverse Aesthetic Goals - Appendix}
%
%
%
%

\author{Xiaoqi~Wang,
        Kevin~Yen,
        Yifan~Hu,
        and~Han-Wei~Shen
\IEEEcompsocitemizethanks{\IEEEcompsocthanksitem Xiaoqi Wang and Han-Wei Shen are with The Ohio State University\protect\\
E-mail: wang.5502@osu.edu, shen.94@osu.edu
\IEEEcompsocthanksitem Kevin Yen and Yifan Hu are with Yahoo! Research\protect\\
Email: kevinyen@yahooinc.com, yifanh@gmail.com}
}

%
%

\markboth{Journal of \LaTeX\ Class Files,~Vol.~14, No.~8, August~2022}%
{Shell \MakeLowercase{\textit{et al.}}: Bare Demo of IEEEtran.cls for Computer Society Journals}
\maketitle
\pagenumbering{gobble}
\IEEEdisplaynontitleabstractindextext

%
\IEEEpeerreviewmaketitle


\section{The Replacement Pattern of \ganName}
\begin{figure}[htbp!]
    \setlength{\abovecaptionskip}{-1pt}
    \begin{center}
        \includegraphics[width=0.9\linewidth]{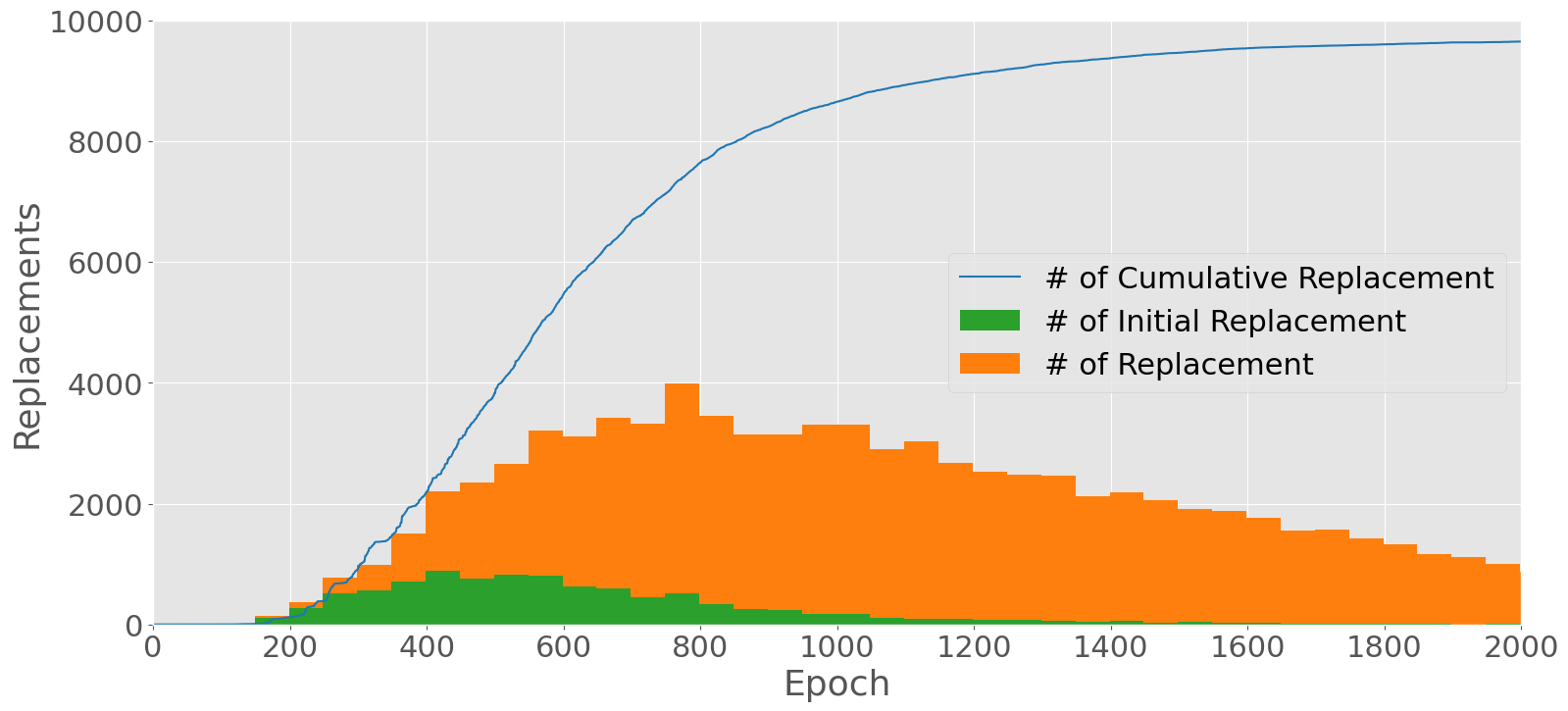}
    \end{center}
    \caption{The number of replacements out of 10,000 training examples over training epochs for \modelName[Xing]. The orange bar represents the number of training graphs whose current good layout examples are replaced. The green bar represents the number of training graphs whose good layout examples are the first time to be replaced by the generated layouts. The blue line shows how many initial good layout examples are replaced cumulatively.}
    \label{fig:replace}
\end{figure}

Lastly, the essence of \ganName\ is continuously improving the good layout examples during the training, which is accomplished by replacing the good layout examples with the better layout generated by itself. For \modelName[Xing], the replacement pattern is shown in \autoref{fig:replace}. As we can see, at the early stage of training, many initial good layout examples were replaced by the generated layouts. After about 1000 epochs, \modelName\ kept replacing the layout generated by itself instead of the initial good layouts. Cumulatively, until the \modelName[Xing] converged, there were more than 9000 training graphs whose initial good layout examples were replaced by the layouts generated by \modelName[Xing].

\section{Computation of Symmetric Percentage Change}
The symmetric percent change (SPC) of a graph $G$ ranging from $-100\%$ to $100\%$ is computed as 
\vspace{-3pt}
\begin{equation}
    \label{equ:spc}
    \mathrm{SPC}_{\lambda}(G) = 100\% \times \frac{\lambda(\mathbf{X}_f,G)-\lambda(\mathbf{X}_b,G)}{\max\{\lambda(\mathbf{X}_f,G),\lambda(\mathbf{X}_b,G)\}}, \\
\vspace{-3pt}
\end{equation}
where $\mathbf{X}_f$ and $\mathbf{X}_b$ are the layouts generated by \modelName\ and a benchmark algorithm respectively. The lower the criterion $\lambda$, the better the layout. The SPC value measures the percentage of difference of $\lambda(\mathbf{X}_b,G)$ compared to $\lambda(\mathbf{X}_f,G)$. To evaluate the relative performance of \modelName\ compared with a benchmark algorithm, the average test SPC is computed as 
\begin{equation}
    \label{equ:average-spc}
    \text{Average Test }\mathrm{SPC}_{\lambda} =  \frac{1}{N_t}\sum_{i=0}^{N_t} \mathrm{SPC}_{\lambda}(G_{i}), \\
\end{equation}
where $N_t$ stands for the total number of graphs in our test set, and $G_i$ is the $i^{th}$ test graph. In terms of the criterion $\lambda$, the average test SPC measures the percentages of the benchmark algorithm outperforms \modelName\ on average.

\section{Equation of Aesthetic Criteria}

\subsection{Stress Majorization}
Stress majorization is the most common objective for graph drawing. It optimizes a graph by simulating the energy release process in a physical spring system. From another perspective, stress energy also measures the discrepancy of node distances between the graph space and the Euclidean layout space. Minimizing the stress energy can significantly improve the aesthetic quality of a graph.
The stress energy for $G$ is computed as
\vspace{0.2cm}
\begin{equation}
\centering
\label{stress}
L_\text{stress}=
\sum_{u, v \in V, u \neq v}
    w_{uv} \left(\|\mathbf{x}_u - \mathbf{x}_v\| - d_{uv}\right)^2 ,
\end{equation}
where the weighting factor $w_{uv}$ is typically  set  to $1/d_{uv}^2$.

\subsection{Crossings Minimization}
Avoiding crossings is an important goal for graph drawing. Graph layouts with fewer crossings usually result in better clarity and less confusion, which is aesthetically better in a general sense. The loss function for minimizing crossing is defined as the sum of all edge crossings in a graph.

\subsection{Crossing Angle Maximization}
Graphs with larger crossing angles tend to be more visually pleasing. The crossing angles in a graph are measured as the acute angles formed by crossing edges. The loss function for maximizing crossing angles is defined as 
\begin{align}
L_\mathrm{xangle} &= \sum_{(\vec{u},\vec{v}) \in \mathrm{crossings(G)}} \left|\arcsin<\frac{\vec{u}}{\|\vec{u}\|},\frac{\vec{v}}{\|\vec{v}\|}>\right|
\end{align}

\subsection{Incident Angle Maximization}
Graphs with larger incident angles tend to be more visually pleasing. The incident angles in a graph are measured as the sharpest angles formed by incident edges over common vertices\cite{purchase-helen-angle}. The loss function for maximizing incident angles is defined as 
\begin{align}
\label{equ:minimum-angle}
    L_\text{angle} = 
    \sum_{v \in V}
    \sum_{\theta_v^{(i)}\in \text{angles}(v)} \bigg|\frac{2\pi}{\deg(v)}-\theta_v^{(i)}\bigg|,
\end{align}
where angle($v$) is the incident angles over node $v$ and deg($v$) denotes the node degree of $v$.

\subsection{Node Occlusion}
Haleem et al. \cite{haleem-huamin} define node occlusion as the total pairs of nodes closer than a threshold. It measures the degree of node clustering in a graph layout. Less local clustering usually leads to more discernible topology structures. Following \cite{deepgd}, we adopt a smooth version of node occlusion as an objective function,
\begin{align}\label{equation:node-occlusion}
L_\text{node occlusion} =
\sum_{\substack{u,v \in V \\ u \neq v}}
e^{-\|\mathbf{x}_u - \mathbf{x}_v\|}.
\end{align}

\subsection{Edge Uniformity}
The uniformity of edges plays a key role in the aesthetic quality of a graph layout.
Aesthetically speaking, a graph layout with more uniform edges is preferable for most cases. One way to measure edge uniformity is to compute the standard deviation of all edge lengths in a graph \cite{haleem-huamin}, 
\begin{align}
L_\text{edge} = 
    \frac{1}{|E|}
    \sum_{(u,v) \in E} \frac{(l_{uv} - \bar{l})^2}{\bar{l}^2},
\qquad \begin{cases}
    l_{uv} = \|\mathbf{x}_u - \mathbf{x}_v\|\\
    \bar{l} = 1\\
\end{cases}
\end{align}
where $l_{uv}$ denotes the edge length between $u$ and v, and $\bar{l}$ is the expected edge length.

\subsection{Shape-Based Metric}
Shape-based metrics, as proposed in previous work \cite{shape-metric}, comprise a family of metrics that assess the quality of a drawing by comparing the similarity between the shape of a set of vertex positions and the original graph topology. The shape of a set of vertex positions can be effectively captured by constructing proximity graphs from these positions. Proximity graph algorithms include Delaunay triangulation, Gabriel graph (GG), relative neighborhood graph (RNG), Euclidean minimum spanning tree (EMST), etc. Within the scope of this work, we adopt the RNG algorithm as the method for obtaining proximity graphs. The similarity between a proximity graph and the original graph can be computed using the mean Jaccard similarity,

\begin{align}
    \mathrm{MJS}(G_\mathrm{prox}, G_\mathrm{orig}) = 
    \frac{1}{|V|}
    \sum_{u \in V}
    \frac{
        |\mathcal{N}_\mathrm{prox}(u) \cap \mathcal{N}_\mathrm{orig}(u)|
    }{
        |\mathcal{N}_\mathrm{prox}(u) \cup \mathcal{N}_\mathrm{orig}(u)|
    },
    \label{equ:mjs}
\end{align}

\noindent where $\mathcal{N}_\mathrm{i}(u)$ denotes the set of neighbours of $u$ in $G_{i}$. To make this aesthetic criterion compatible with our framework, we compute the shape metric score by measuring the distance instead of similarity as follows,

\begin{align}
    L_\mathrm{shape} = 1 - \mathrm{MJS}(G_\mathrm{prox}, G_\mathrm{orig}).
\end{align}

\subsection{t-SNE Score}
The t-distributed stochastic neighbor embedding (t-SNE) is a dimensionality reduction algorithm widely used in the field of visualization. A previous work called tsNET\cite{tsnet} shows that t-SNE can also be applied to graph visualization by minimizing the t-SNE score of a graph layout, which indicates the divergence between its graph space and its layout space. So, this objective has a similar effect to stress majorization. However, minimizing the t-SNE score will also encourage local clustering. The t-SNE score of a graph layout is defined as follows,

\begin{align}
    L_\text{t-SNE} = \sum_{\substack{i,j\\i \neq j}} p_{ij}\log\frac{p_{ij}}{q_{ij}}.
    \label{equ:tsne-loss}
\end{align}

\noindent where

\begin{align}
    p_{ij} = p_{ji} &= \frac{p_{j|i}+p_{i|j}}{2N} \\ 
    p_{j|i} &= \frac{\exp(-\frac{d_{ij}^2}{2\sigma_i^2})}{\sum_{\substack{k\\k\neq i}}\exp(-\frac{d_{ik}^2}{2\sigma_i^2})} \\
    q_{ij} = q_{ji} &= \frac{(1+||\mathbf{x}_i-\mathbf{x}_j||^2)^{-1} }{\sum_{\substack{k,l\\k \neq l}}(1+||\mathbf{x}_k-\mathbf{x}_l||^2)^{-1}}
\end{align}

%
%
%
%


\section{More Qualitative Comparison}
We managed to conduct a more comprehensive qualitative comparison study by presenting the qualitative comparison of 95 graphs unseen during the model training. These 95 unseen graphs include 38 real-world large graphs from the SuiteSparse Matrix Collection (see Figure~\ref{fig:more-vis-result1} and Figure~\ref{fig:more-vis-result2}) and 57 Rome graphs (see Figure~\ref{fig:more-vis-result3}, Figure~\ref{fig:more-vis-result4} and Figure~\ref{fig:more-vis-result5}).
\newcommand{\imgcell}[1]{\raisebox{-0.4\totalheight}{\adjustbox{height=30px, trim={0.08\width} {0.08\height} {0.08\width} {0.08\height},clip}{\includegraphics[]{#1}}}}

\begin{table*}[ht!]
\setlength{\tabcolsep}{0pt}
\renewcommand{\arraystretch}{0}
\fontsize{6}{6}\selectfont
\centering
\begin{tabular}{ c|ccccc|ccccccc }
    \bfseries{\thead{Graph}} & \multicolumn{5}{c|}{\thead{Benchmark Methods}} & \multicolumn{6}{c}{\thead{SmartGD}}\\
    & \bfseries{SGD2}
    & \bfseries{PMDS}
    & \bfseries{FA2}
    & \bfseries{DeepGD}
    & \makecell{\bfseries GD2\\\relax[Stress+Xing]}
    & \makecell{\bfseries SmartGD\\\relax[Stress]}
    & \makecell{\bfseries SmartGD\\\relax[Xing]}
    & \makecell{\bfseries SmartGD\\\relax[Shape]}
    & \makecell{\bfseries SmartGD\\\relax[XAngle]}
    & \makecell{\bfseries SmartGD\\\relax[Stress+Xing]}
    & \makecell{\bfseries SmartGD\\\relax[Stress+XAngle]}
    & \makecell{\bfseries SmartGD\\\relax[7-Aesthetics]}
    \rule[-1ex]{0pt}{0ex} \\ \hline

\makecell{\bfseries utm300\\N = 300\\M = 2191} &
\imgcell{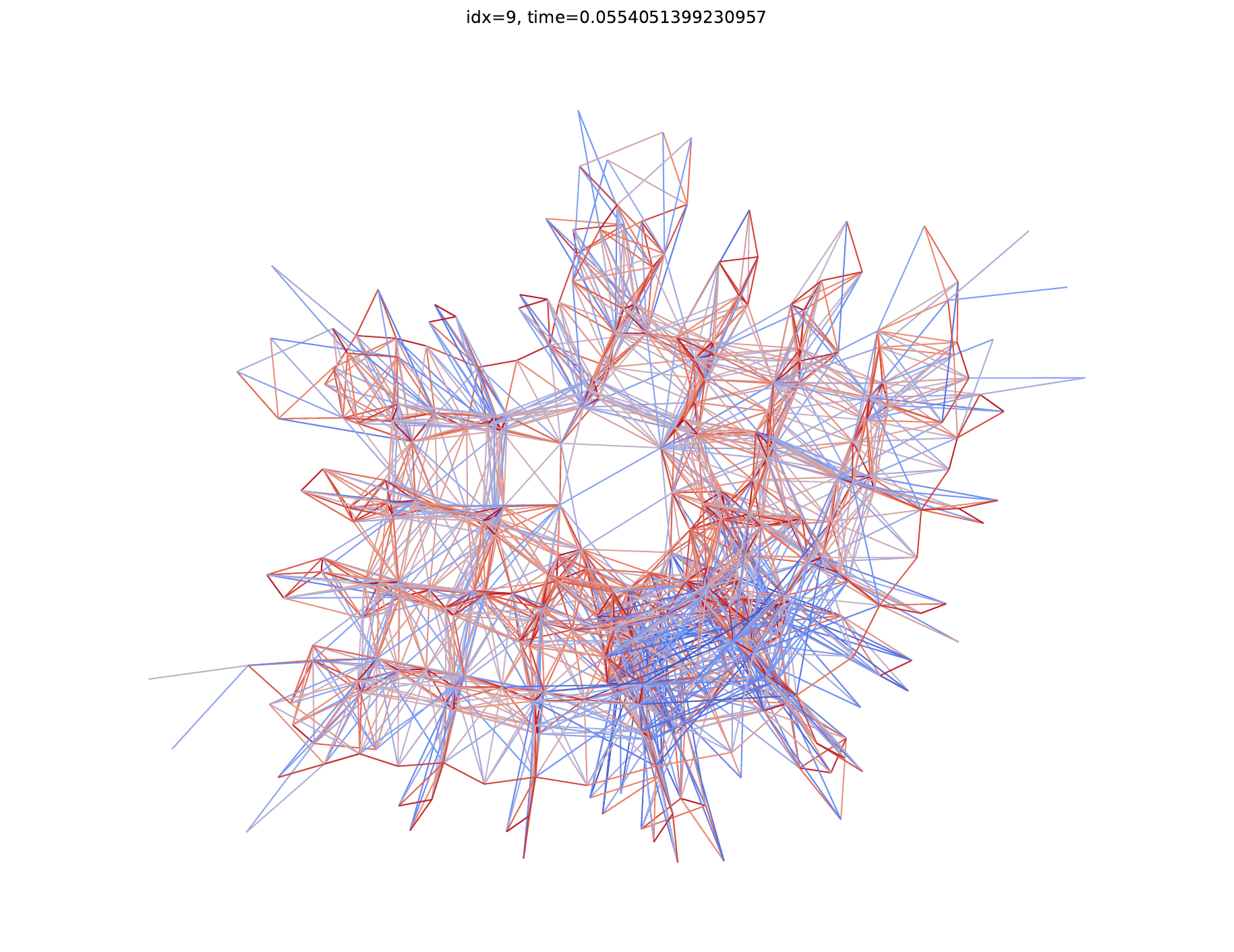} &
\imgcell{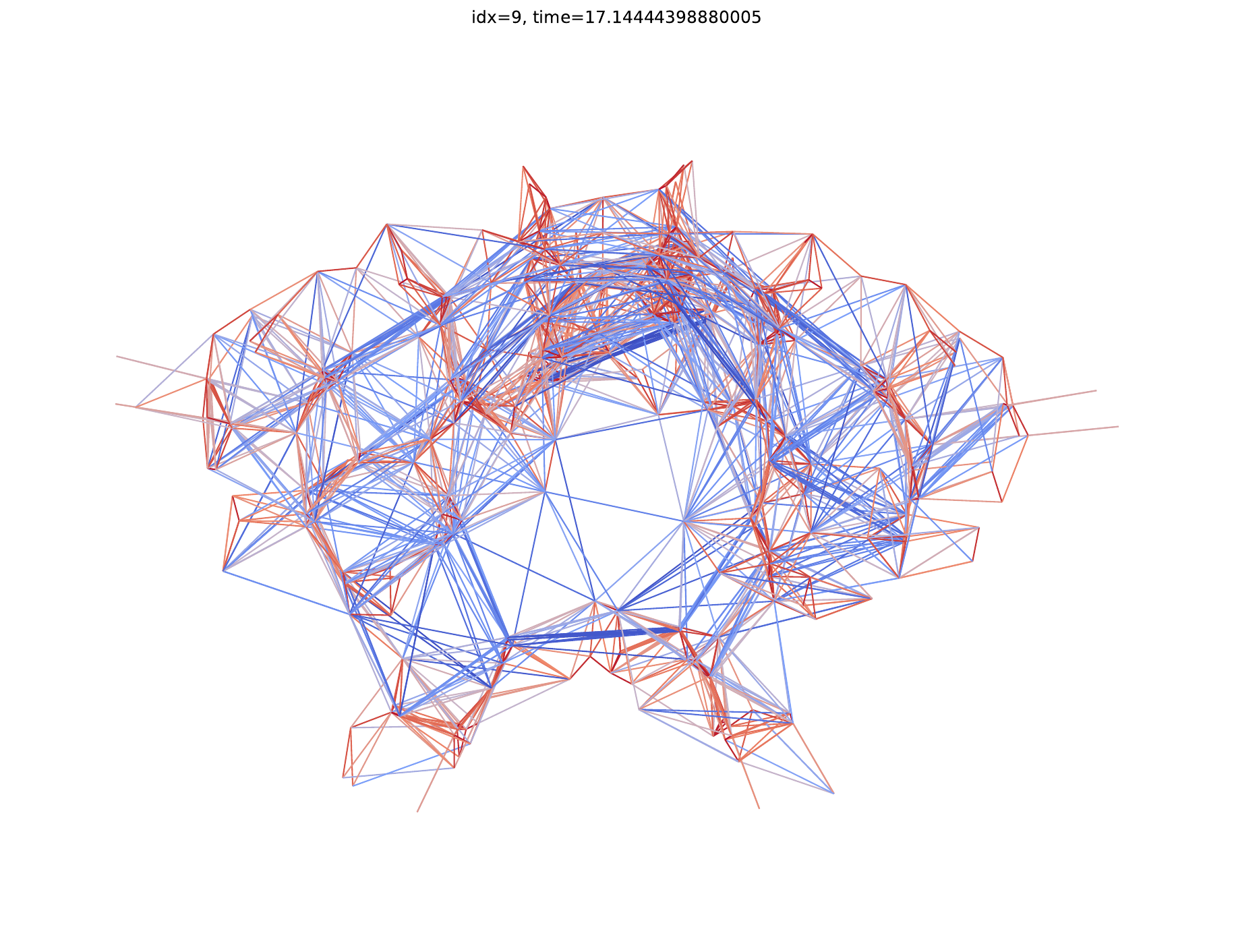} &
\imgcell{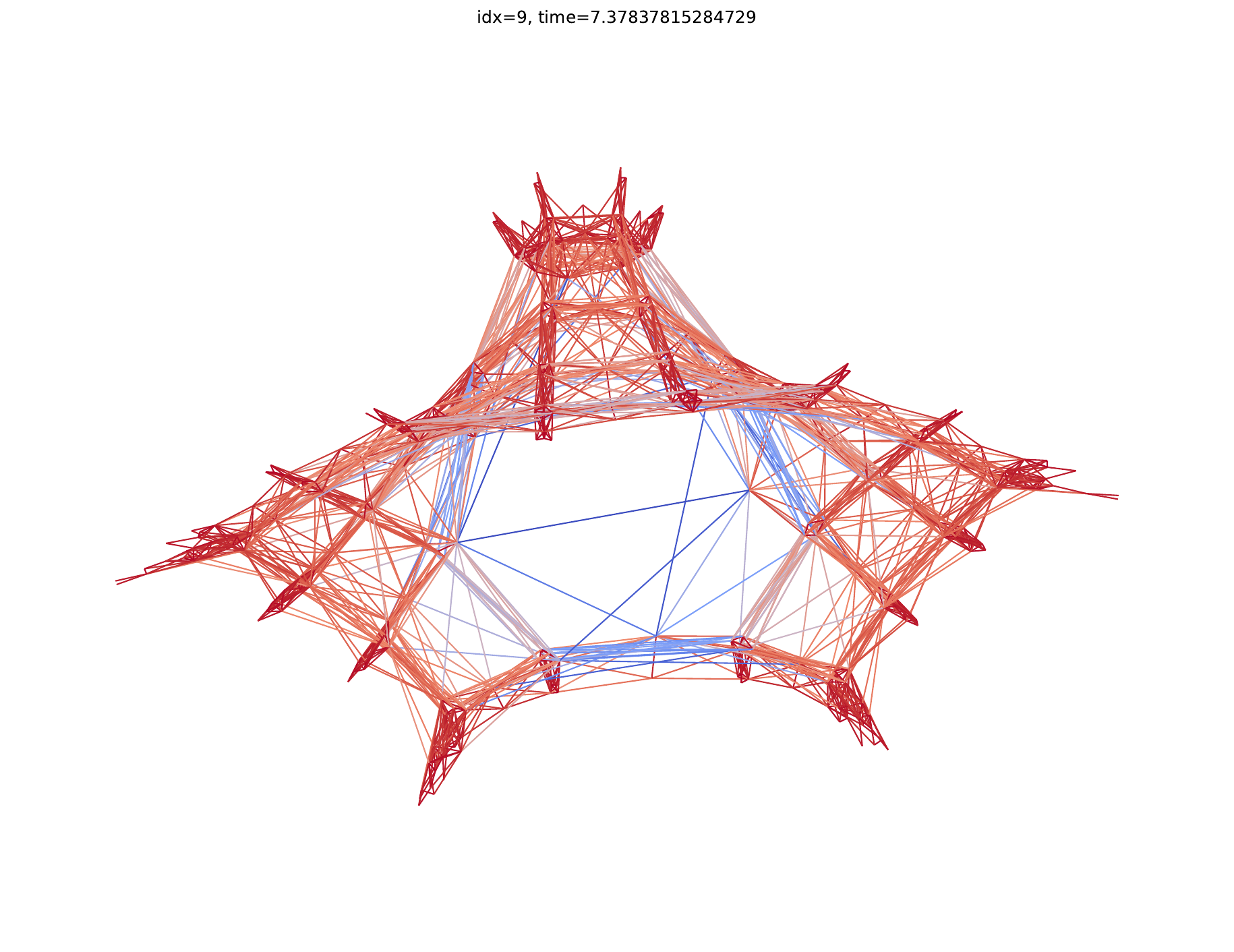} &
\imgcell{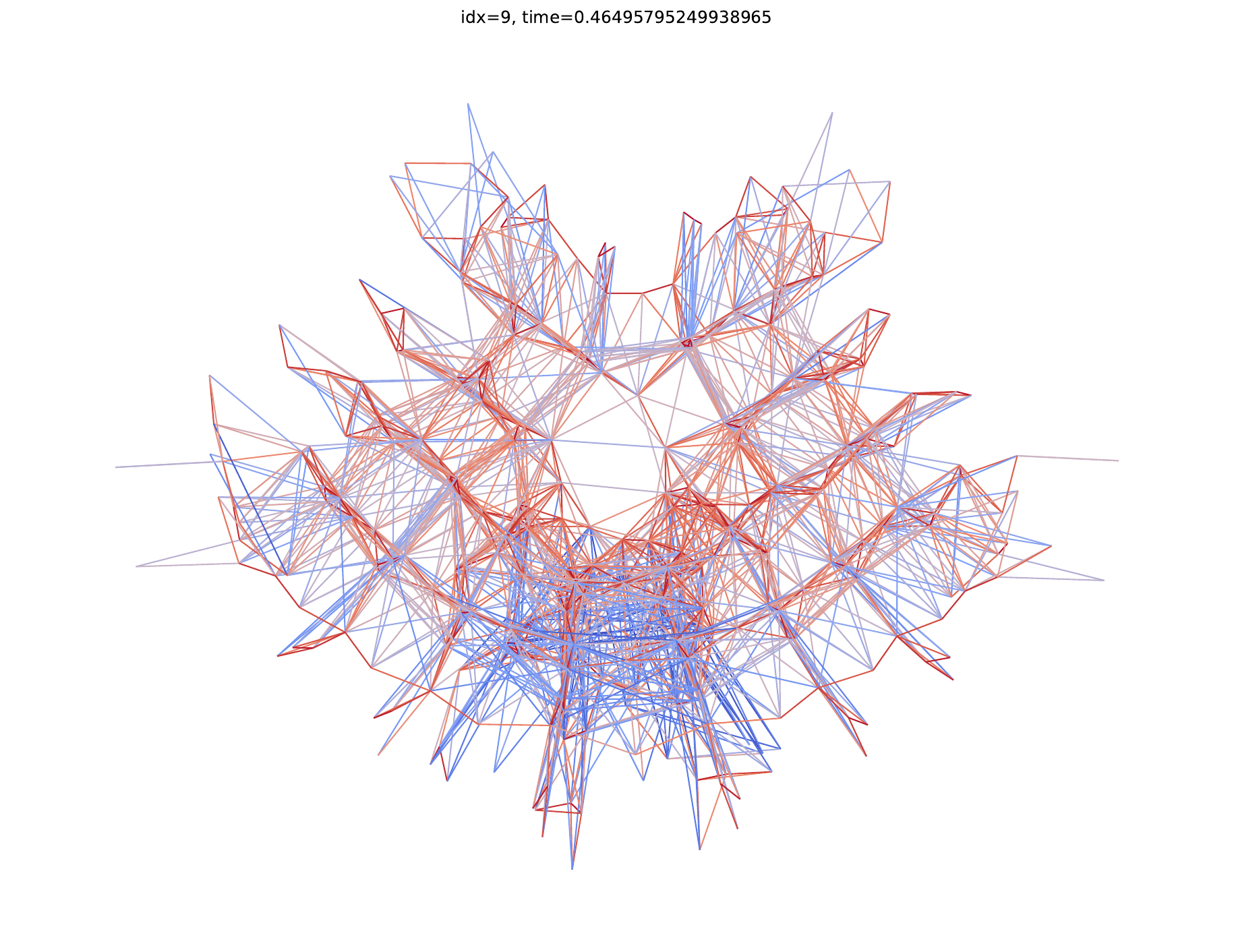} &
\imgcell{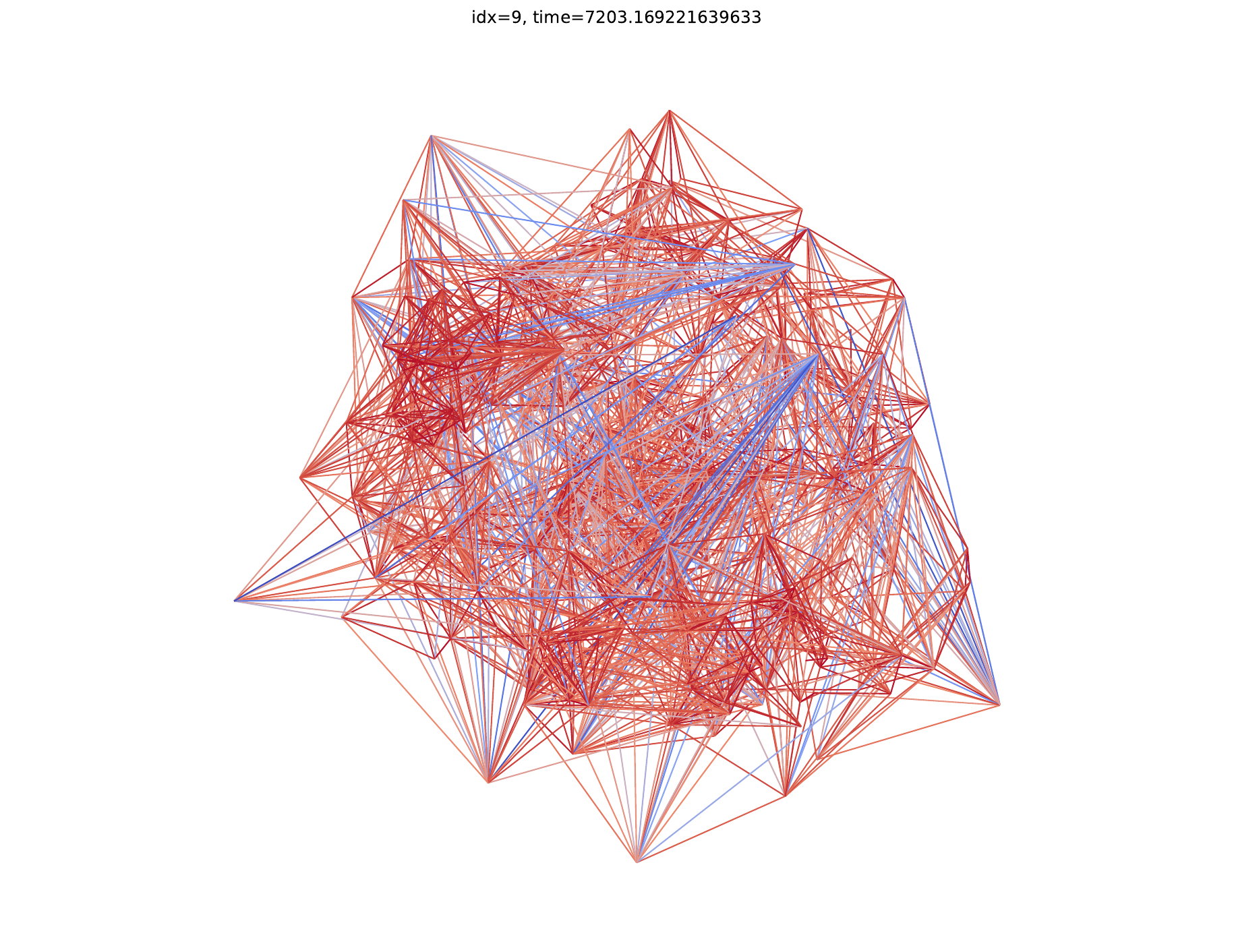} &
\imgcell{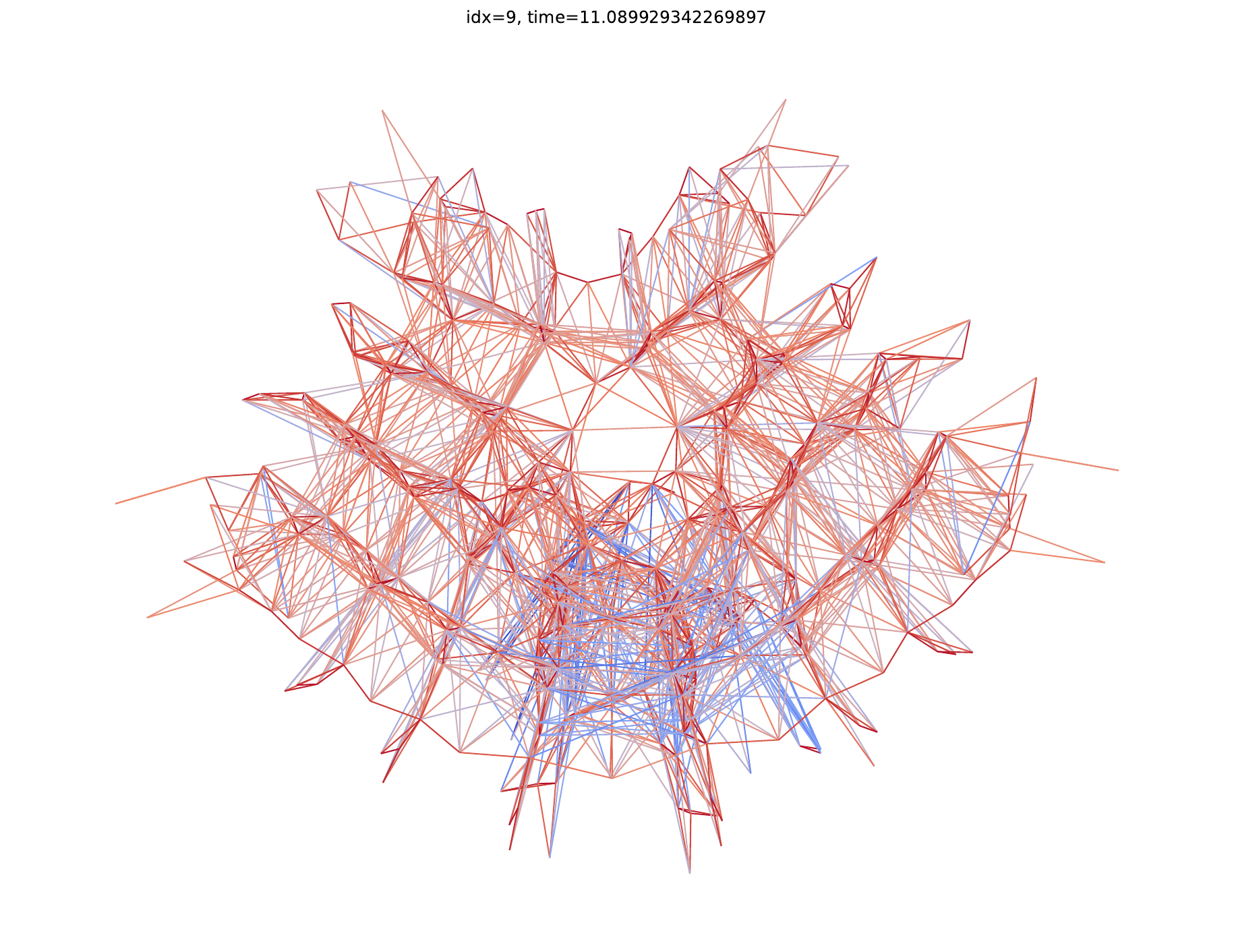} &
\imgcell{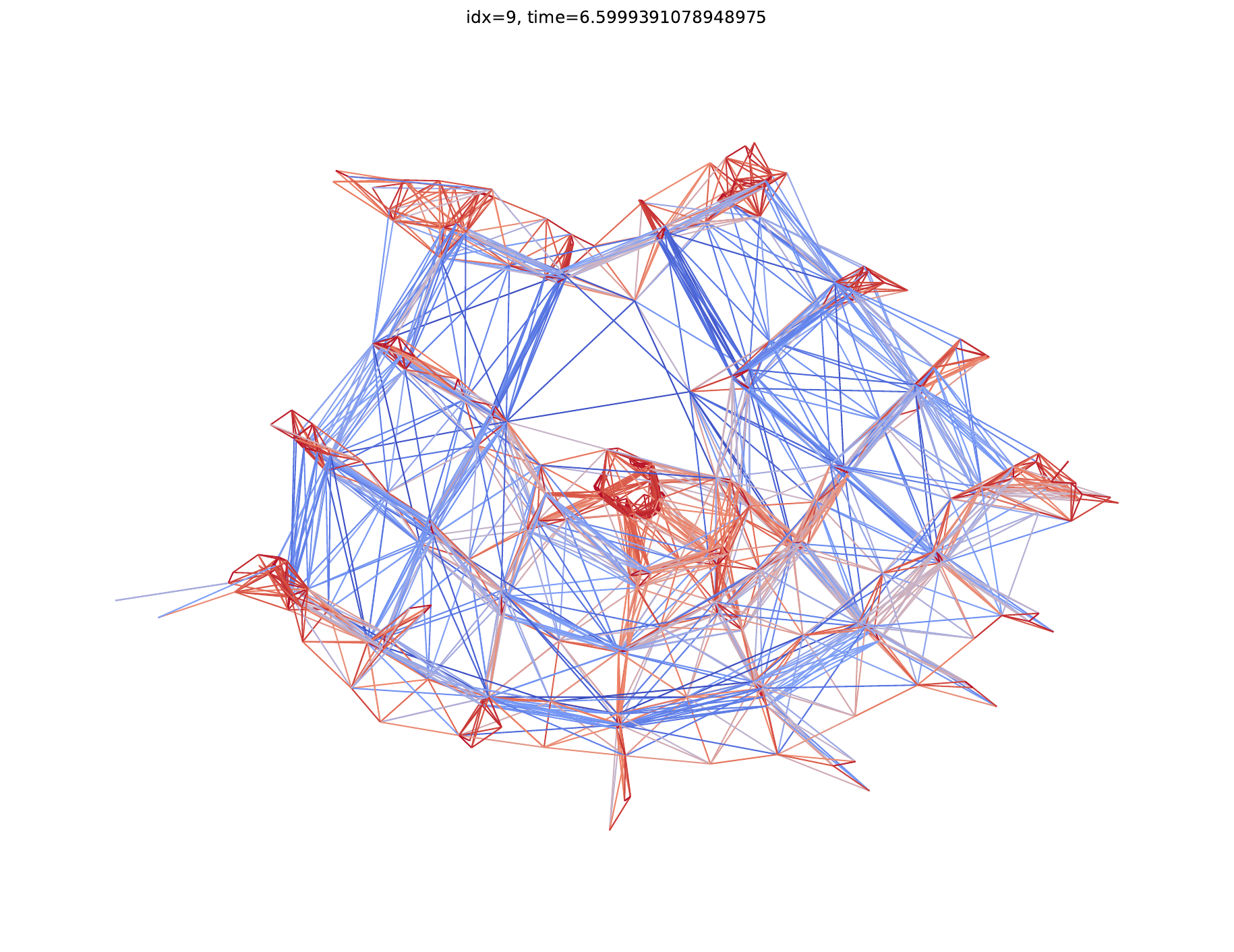} &
\imgcell{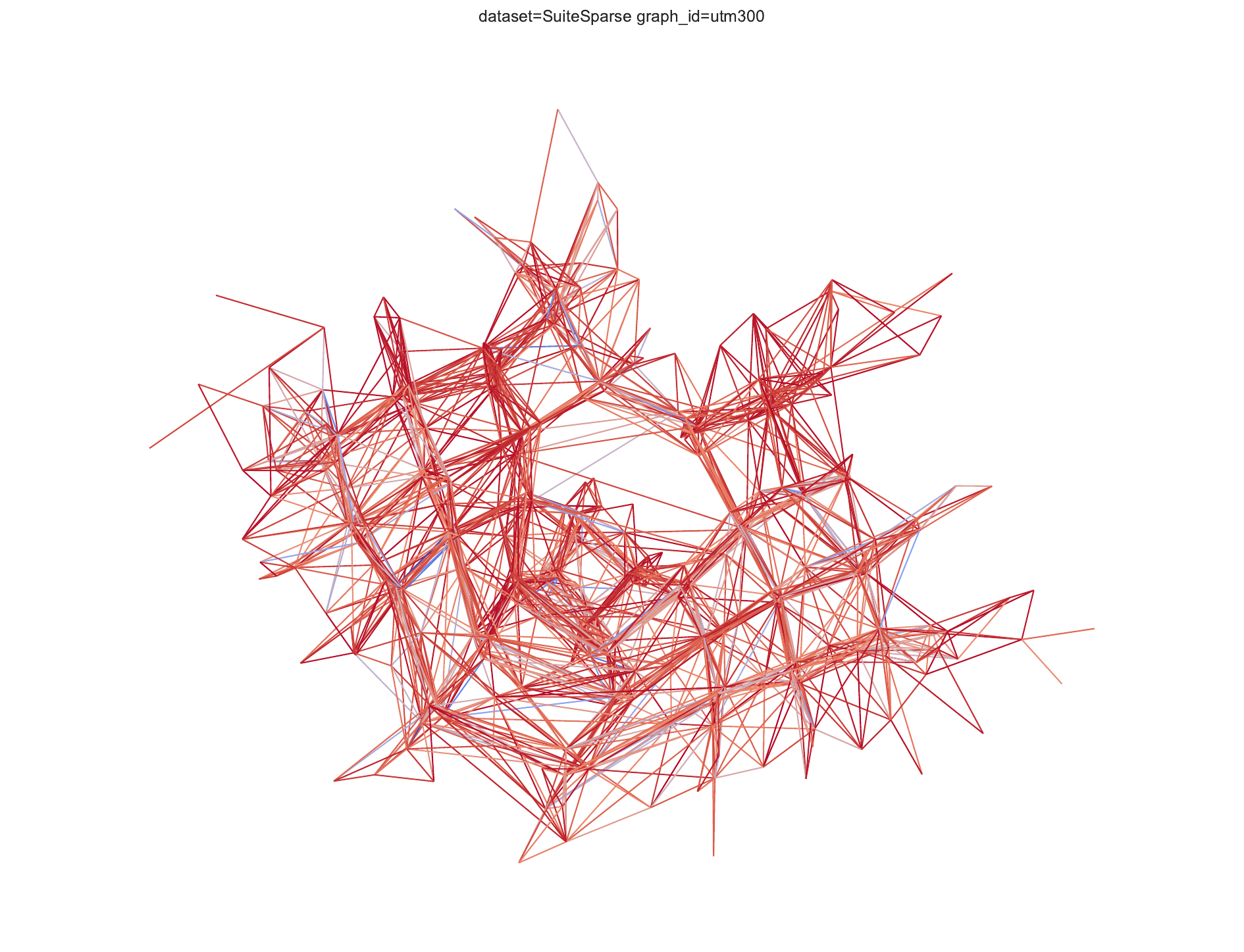} &
\imgcell{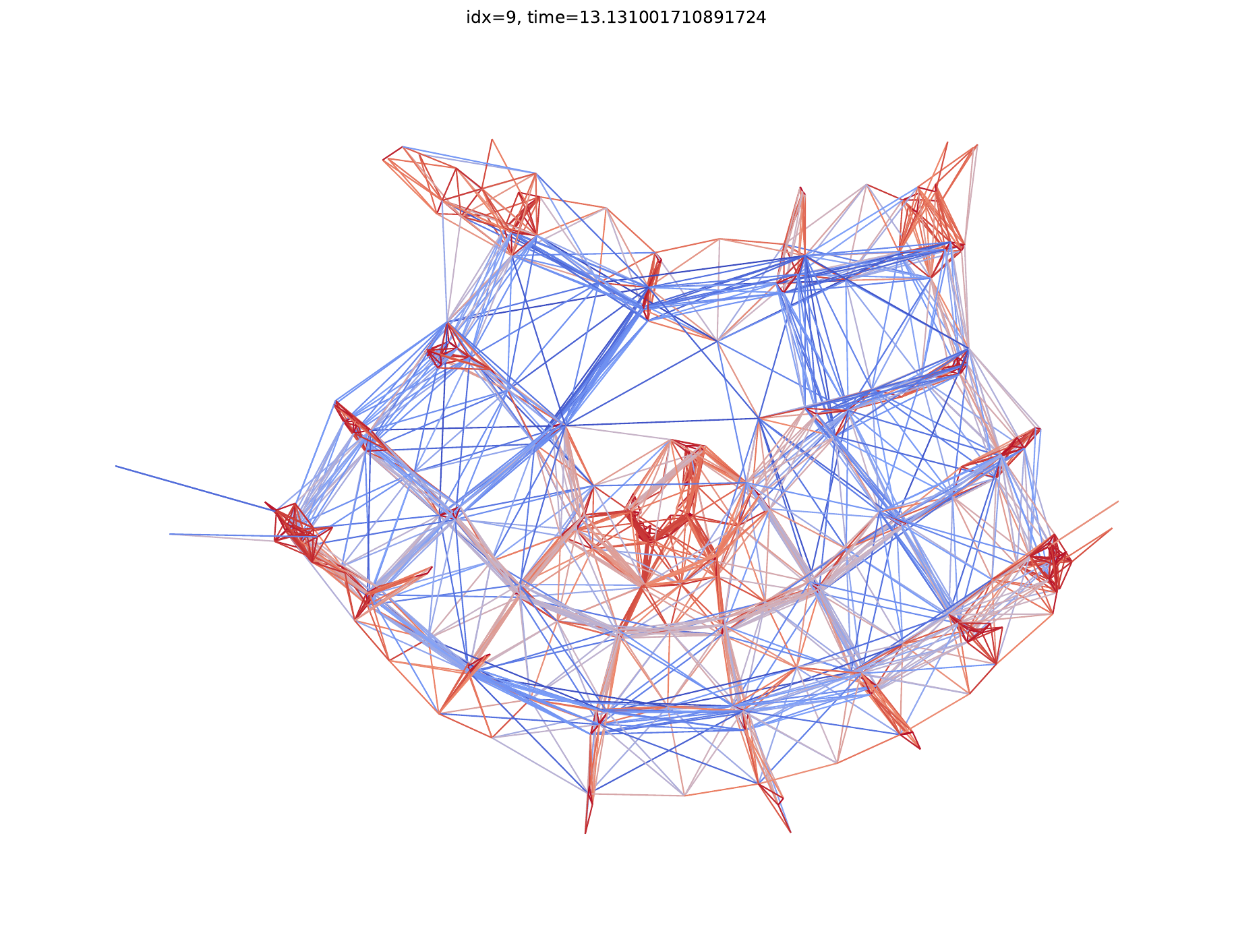} &
\imgcell{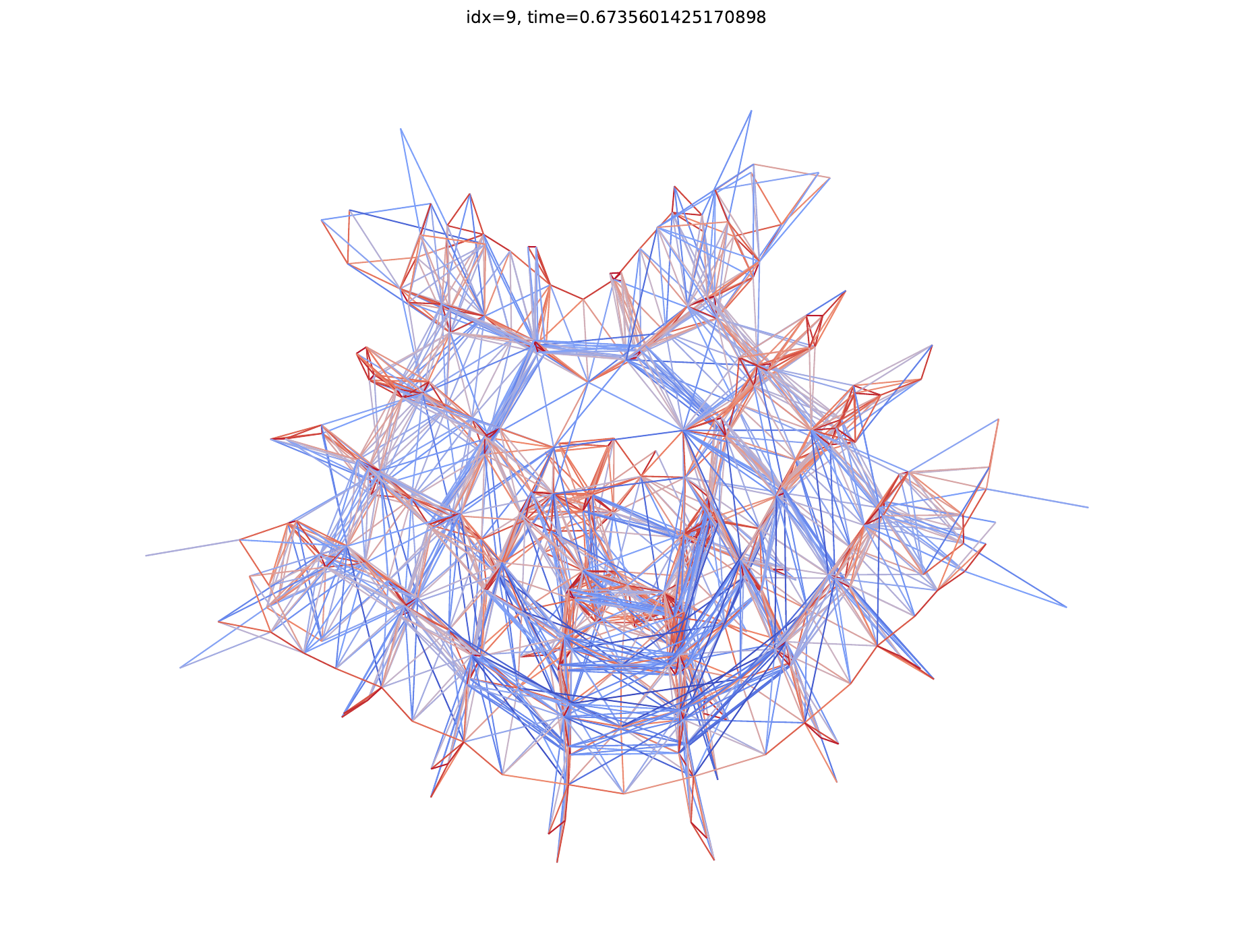} &
\imgcell{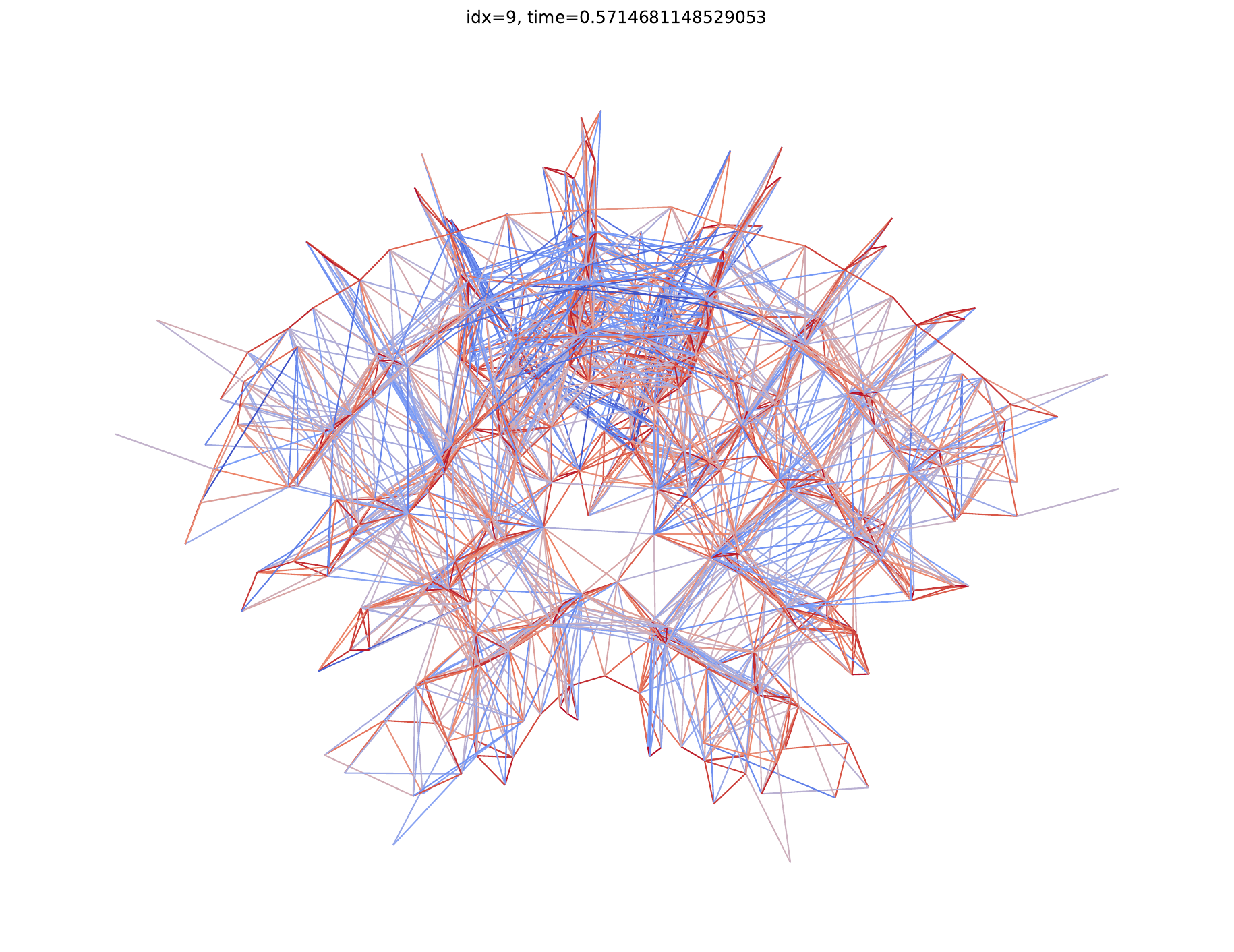} &
\imgcell{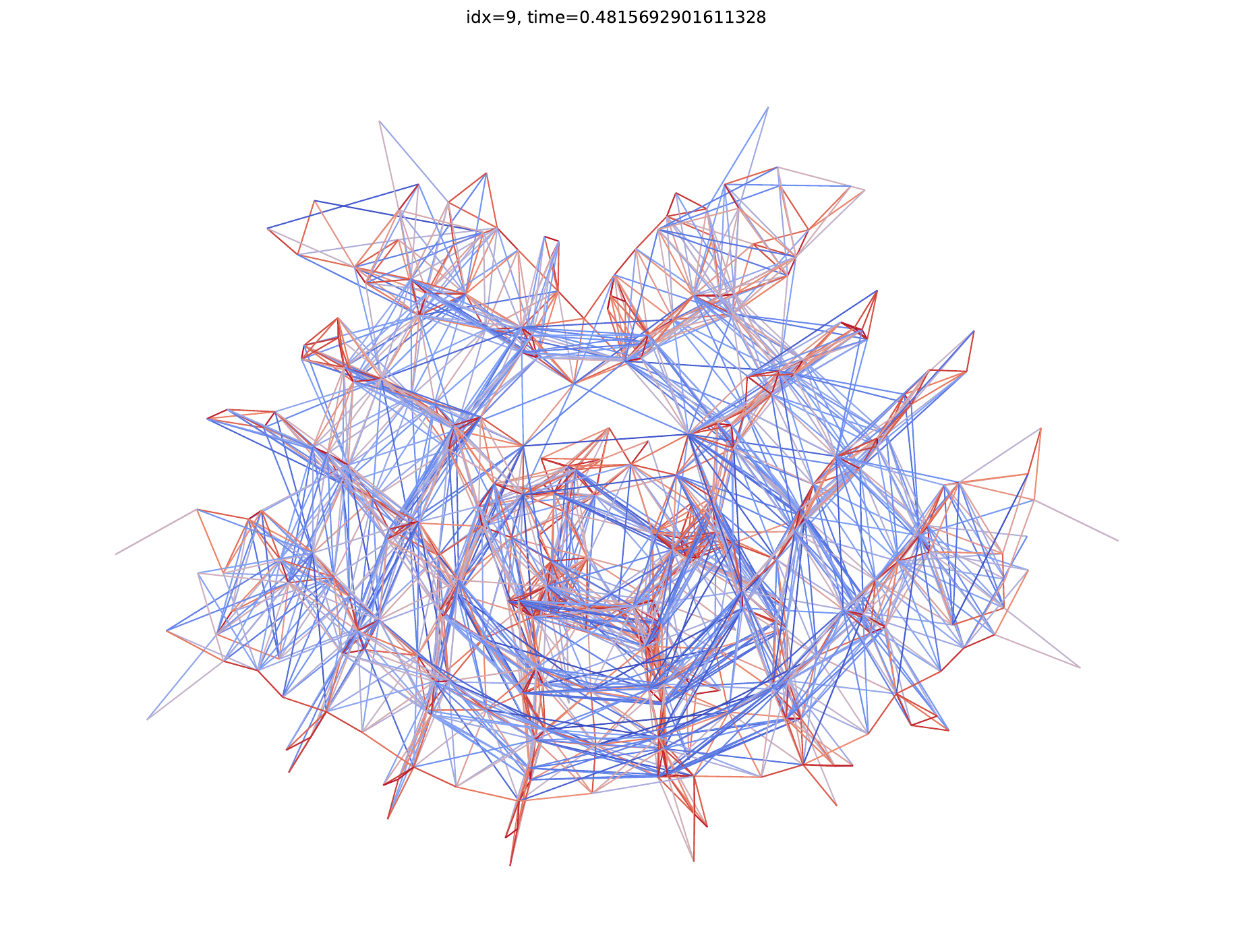} \\

&
t = 0.06s &
t = 17.14s &
t = 7.38s &
t = 0.46s &
t = 7200.00s &
t = 0.56s &
t = 0.31s &
t = 0.77s &
t = 0.56s &
t = 0.70s &
t = 0.68s &
t = 0.53s \\

\makecell{\bfseries mesh2e1\\N = 306\\M = 856} &
\imgcell{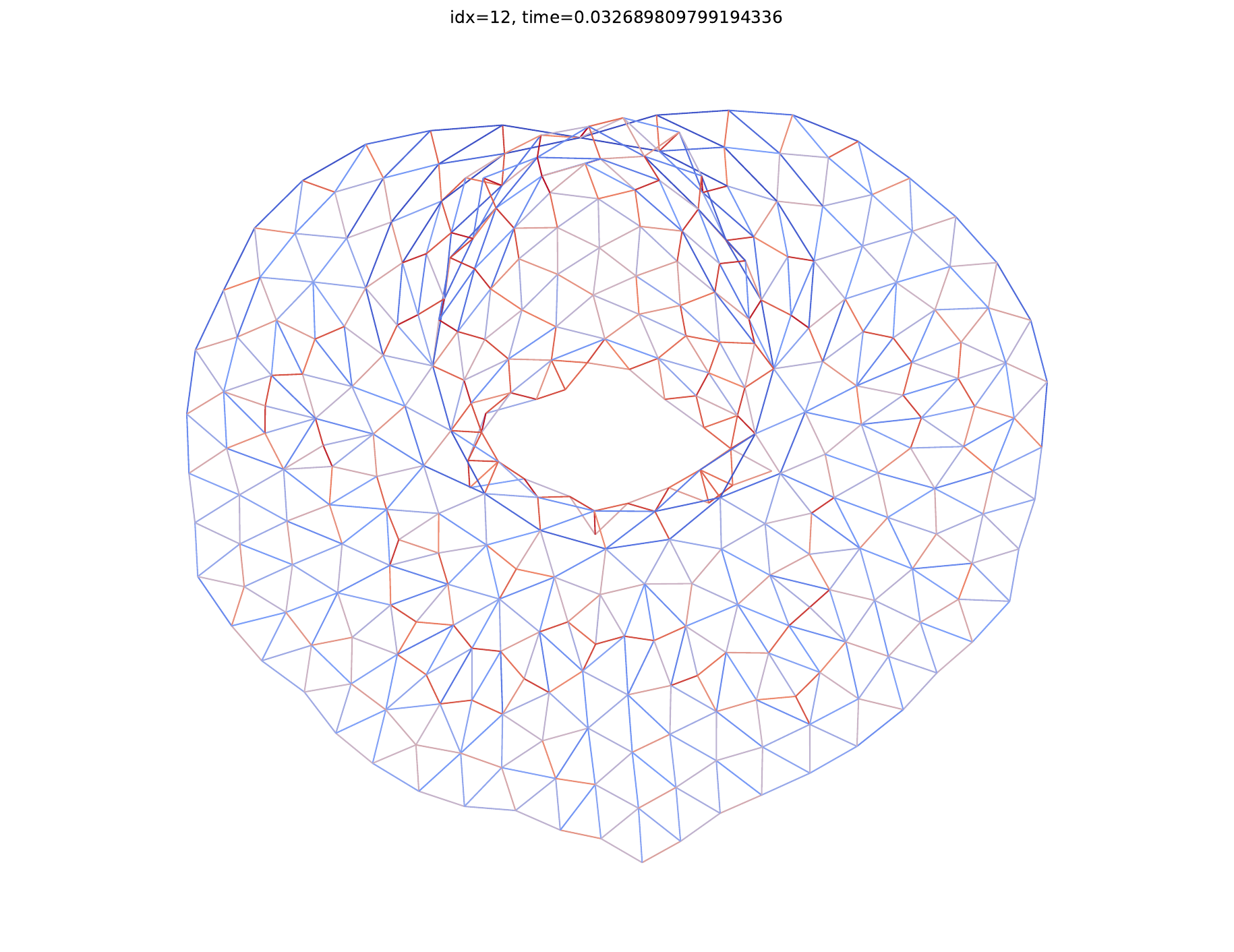} &
\imgcell{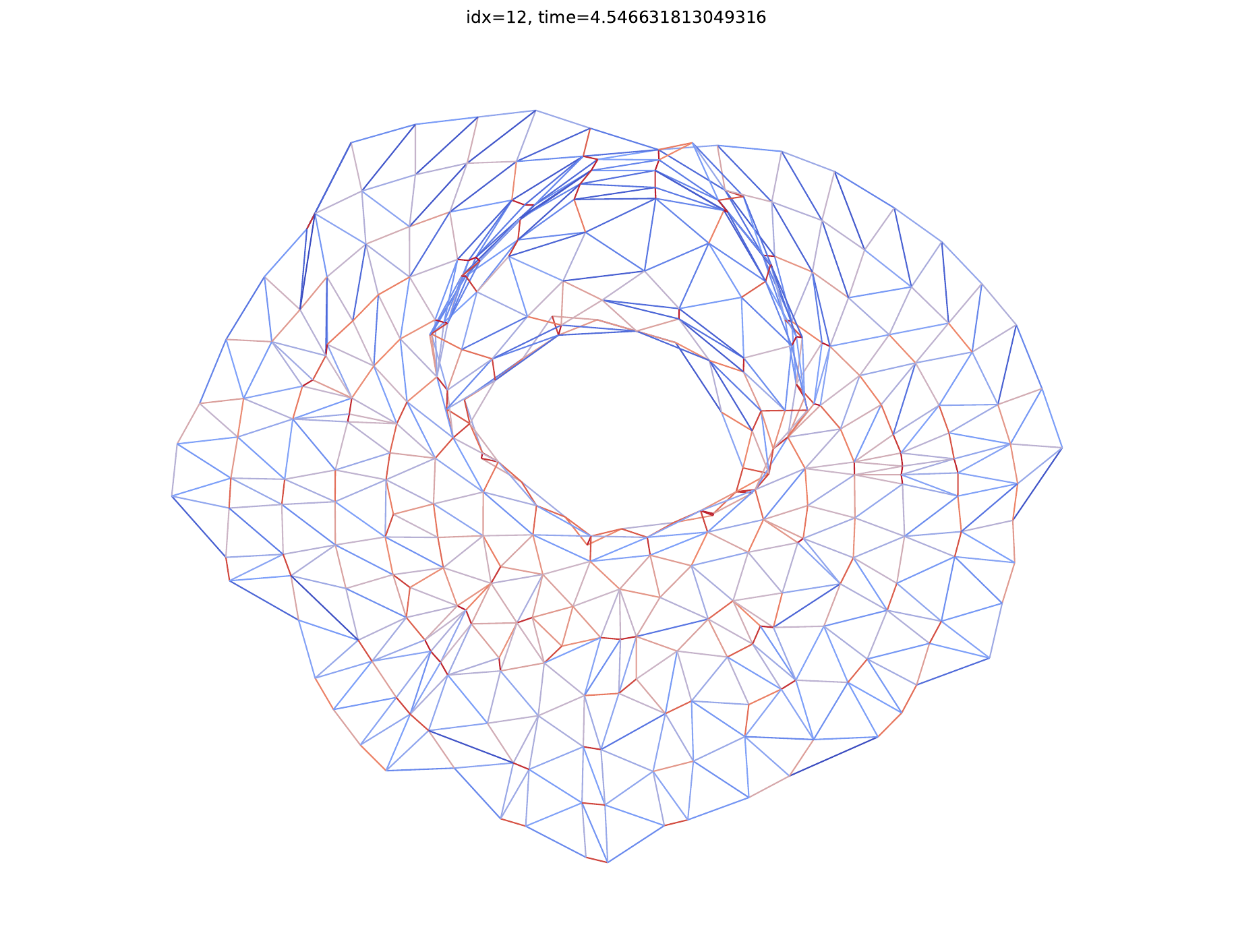} &
\imgcell{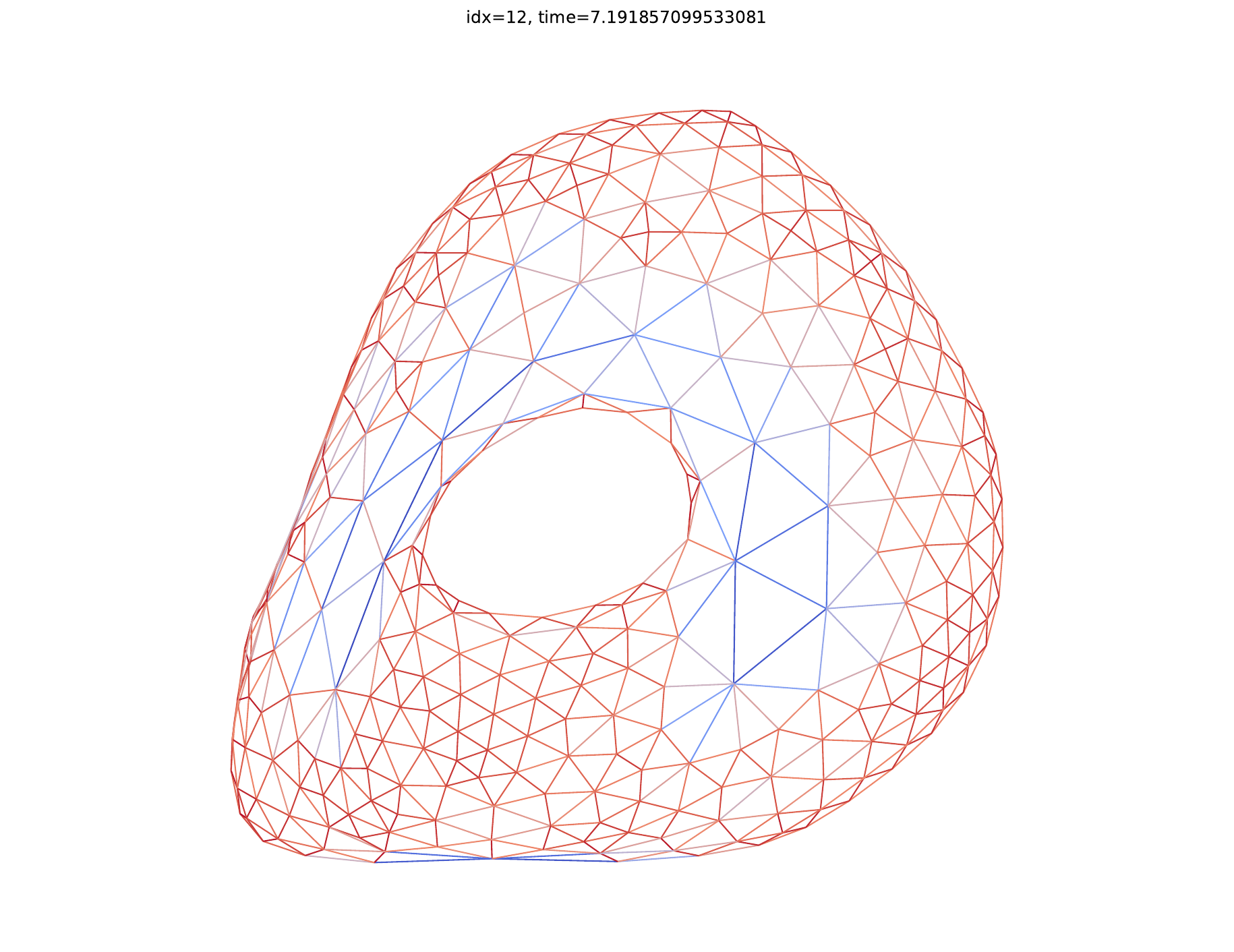} &
\imgcell{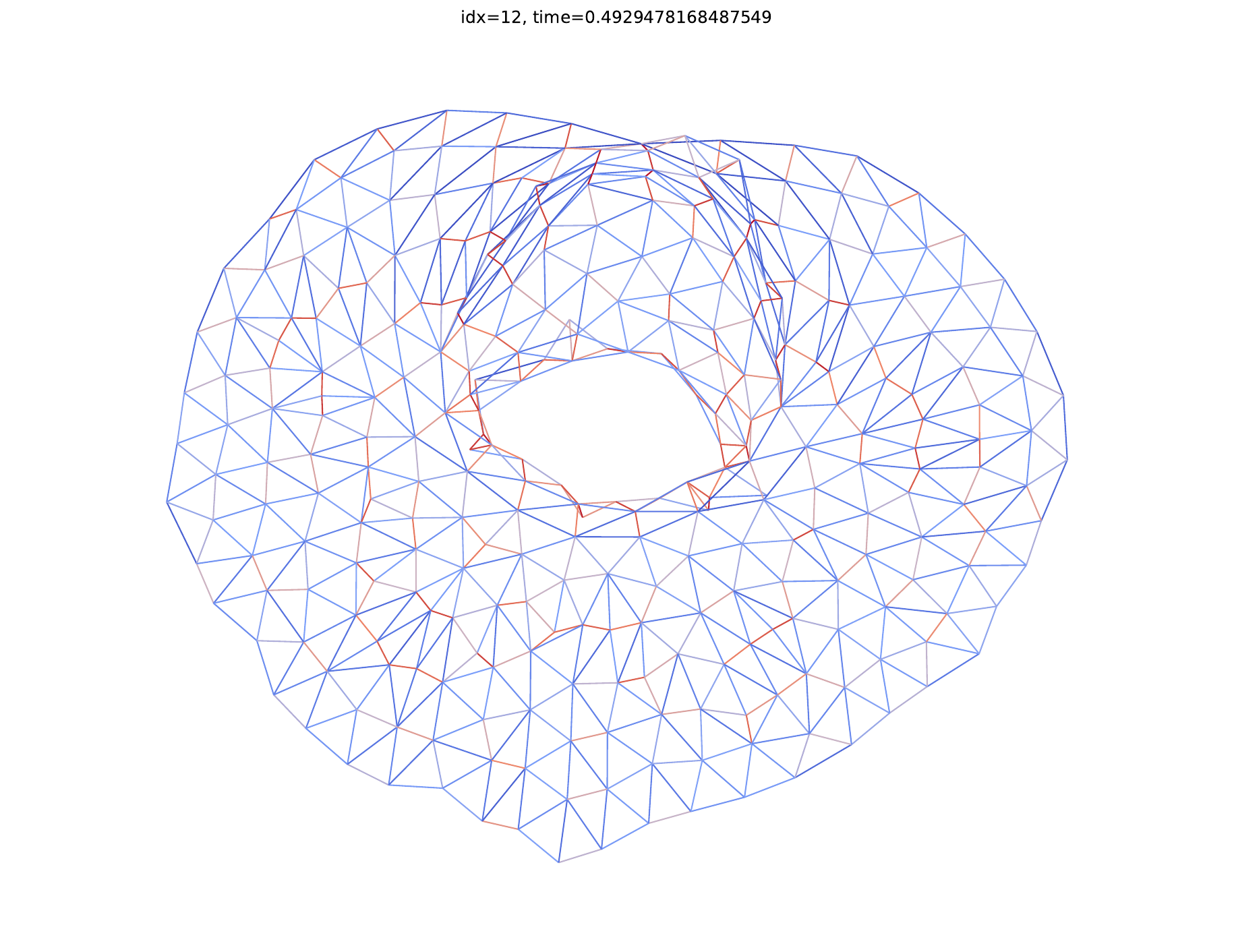} &
\imgcell{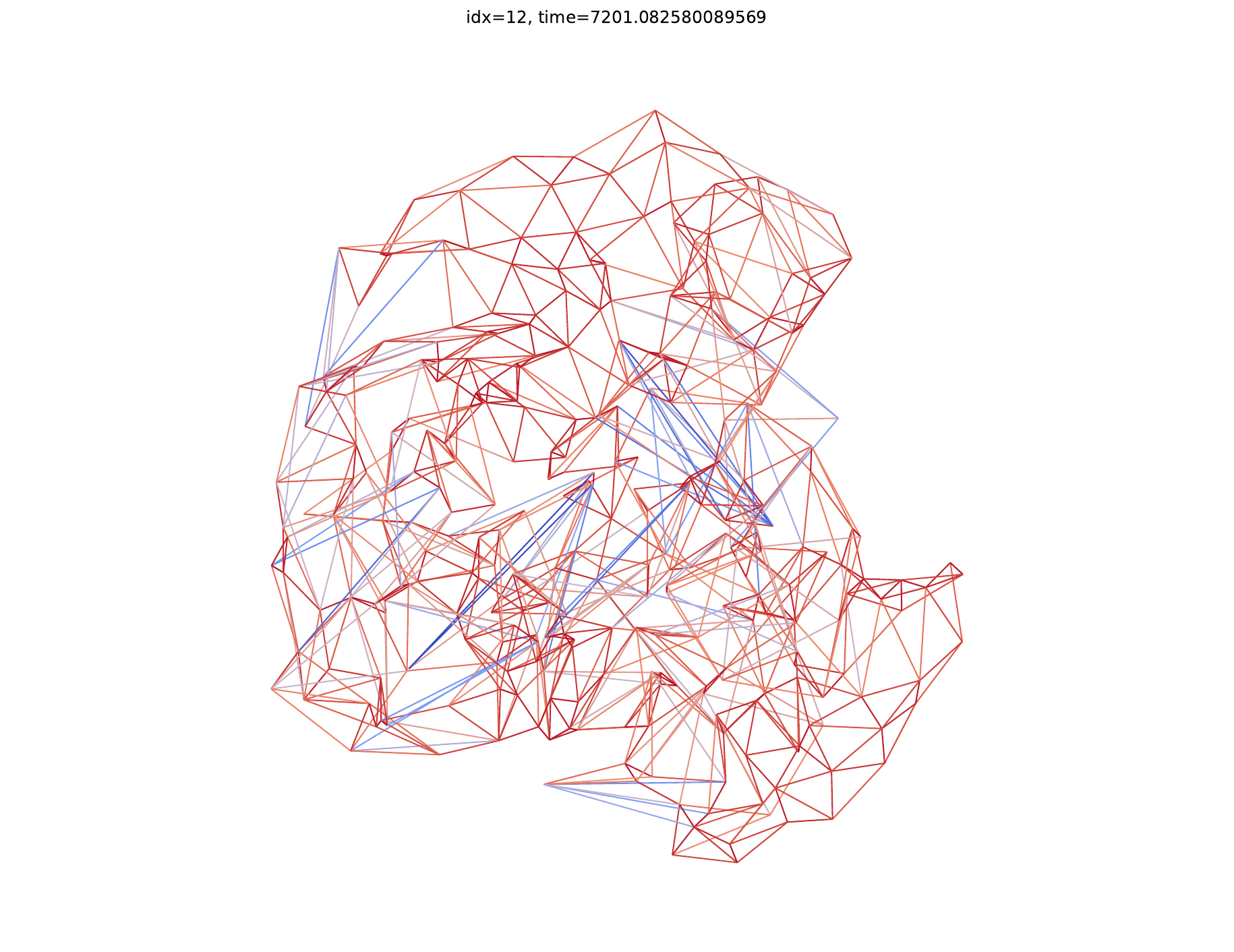} &
\imgcell{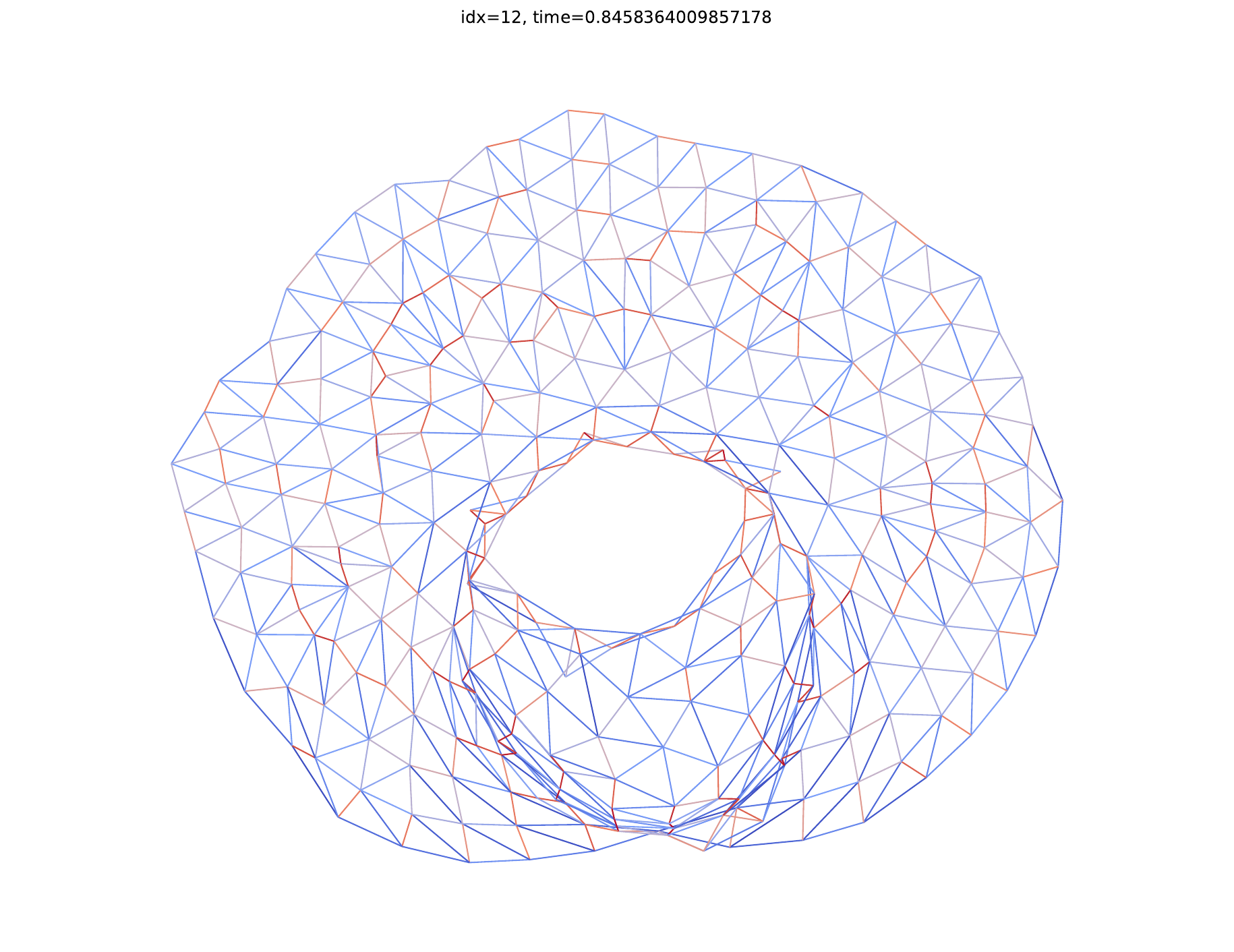} &
\imgcell{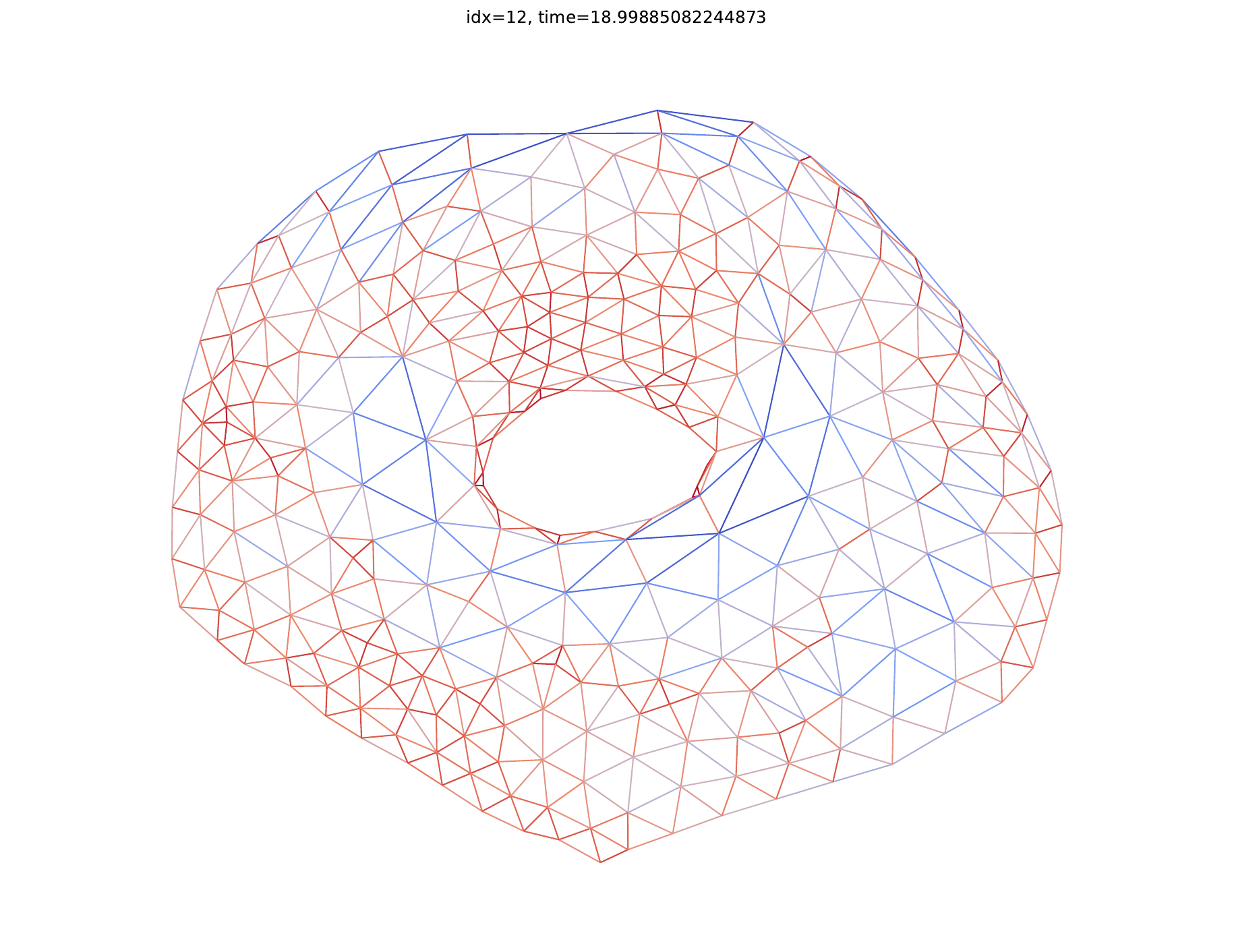} &
\imgcell{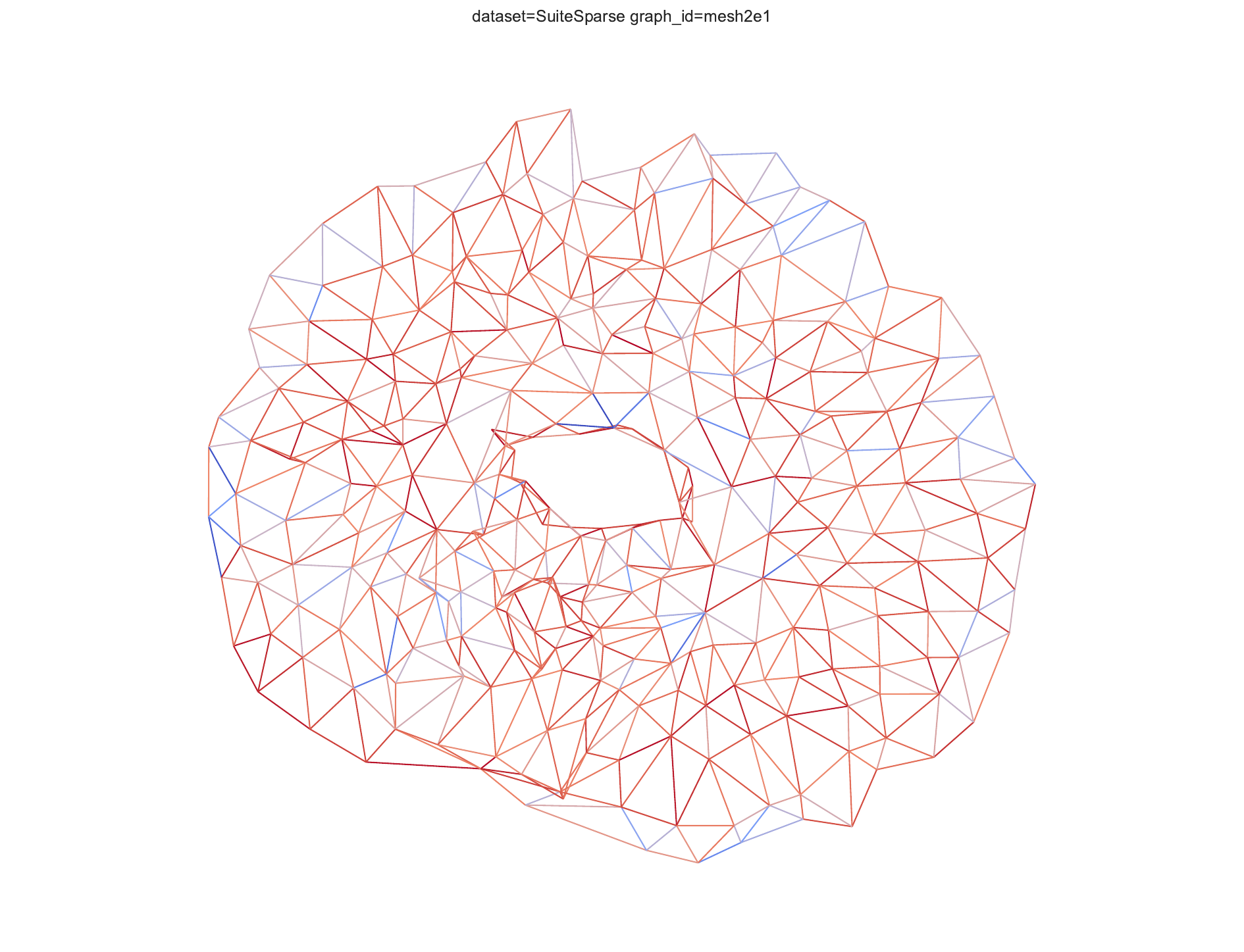} &
\imgcell{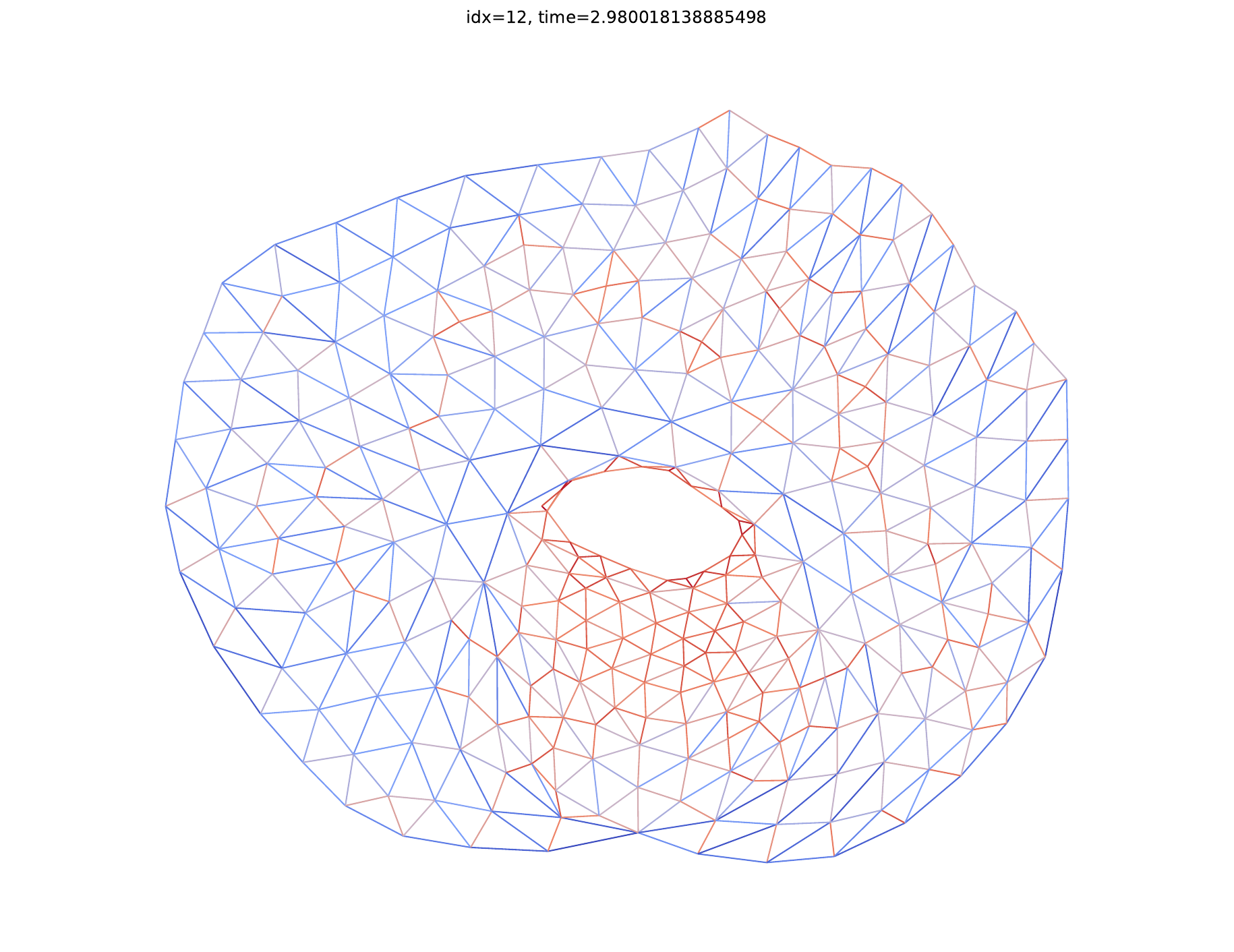} &
\imgcell{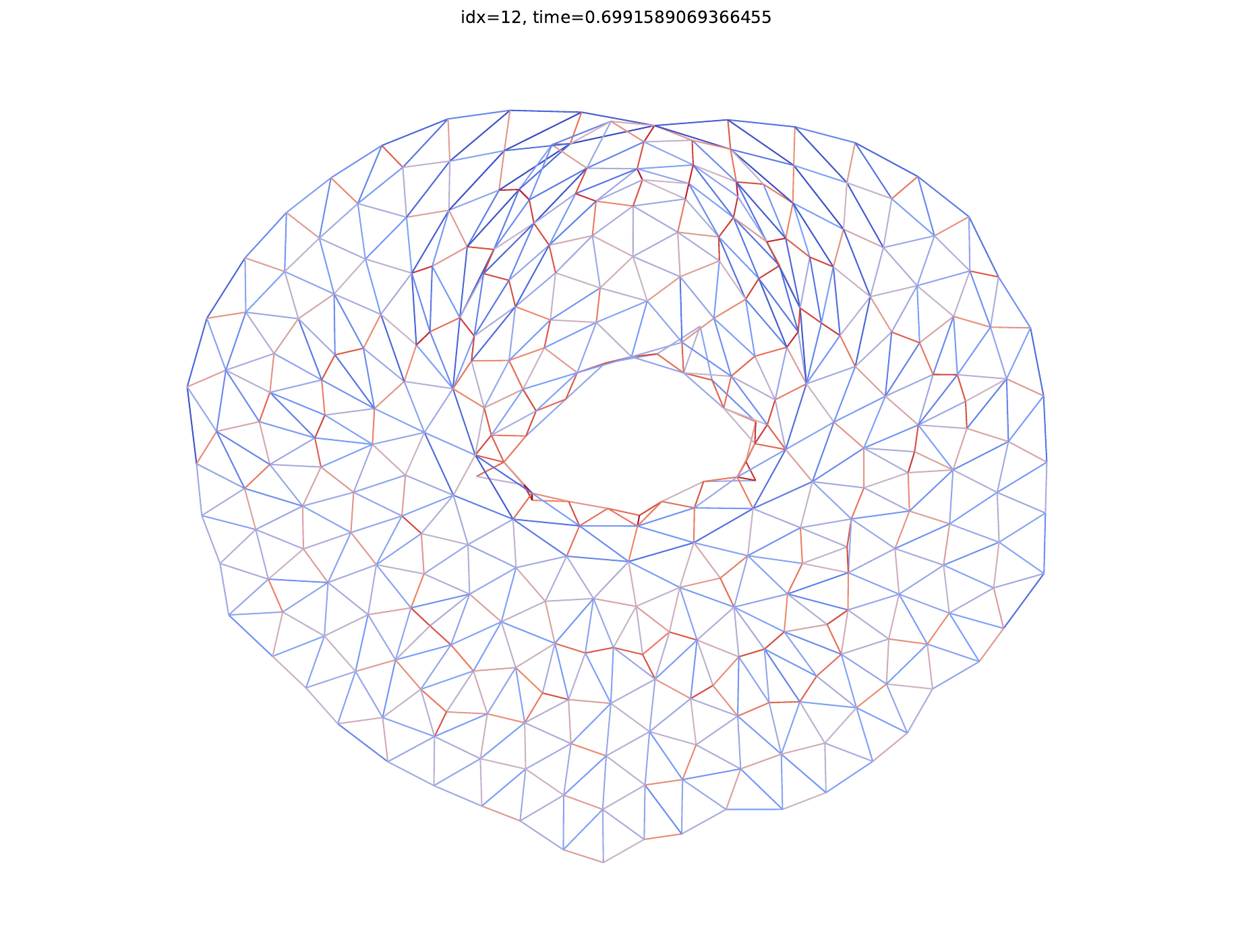} &
\imgcell{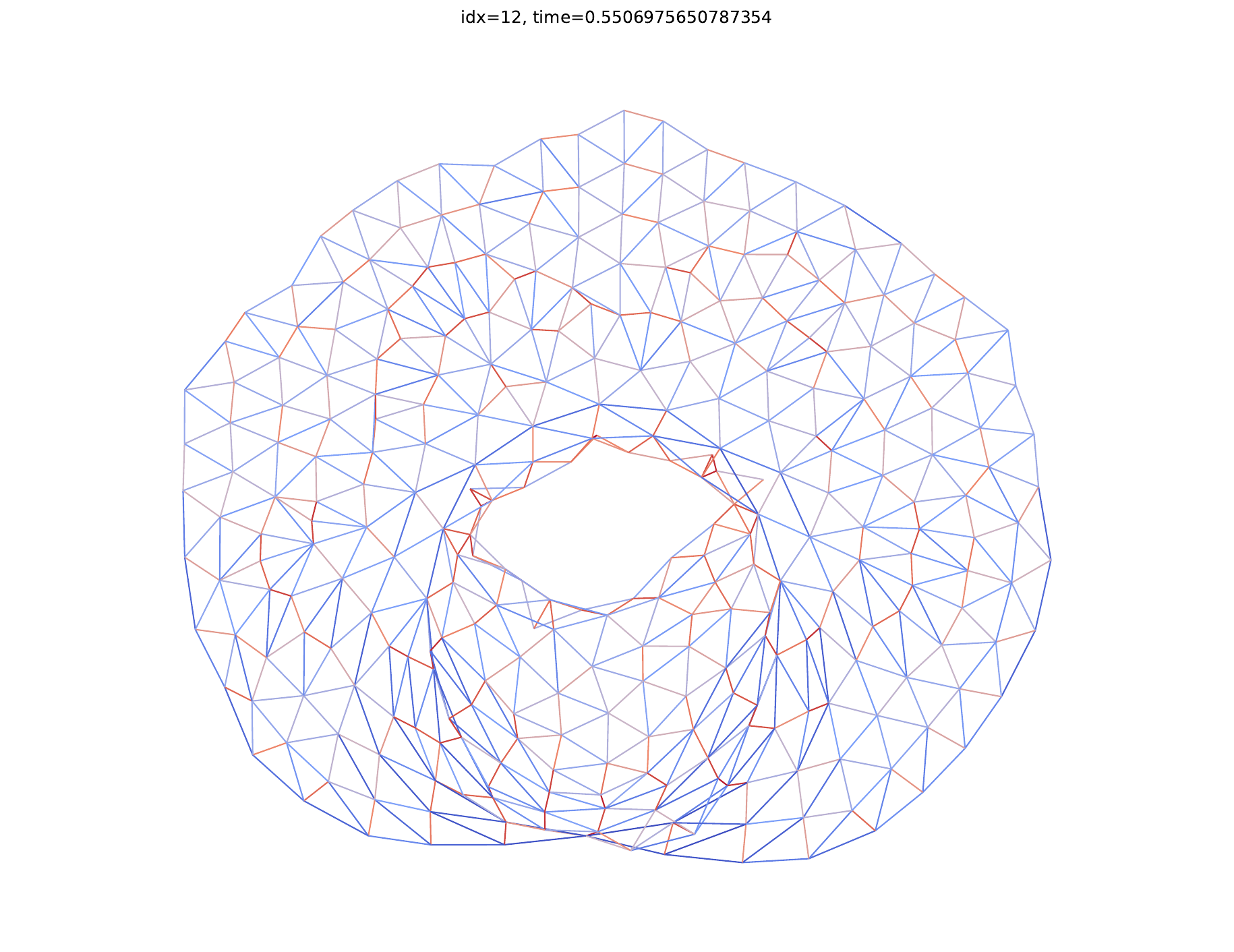} &
\imgcell{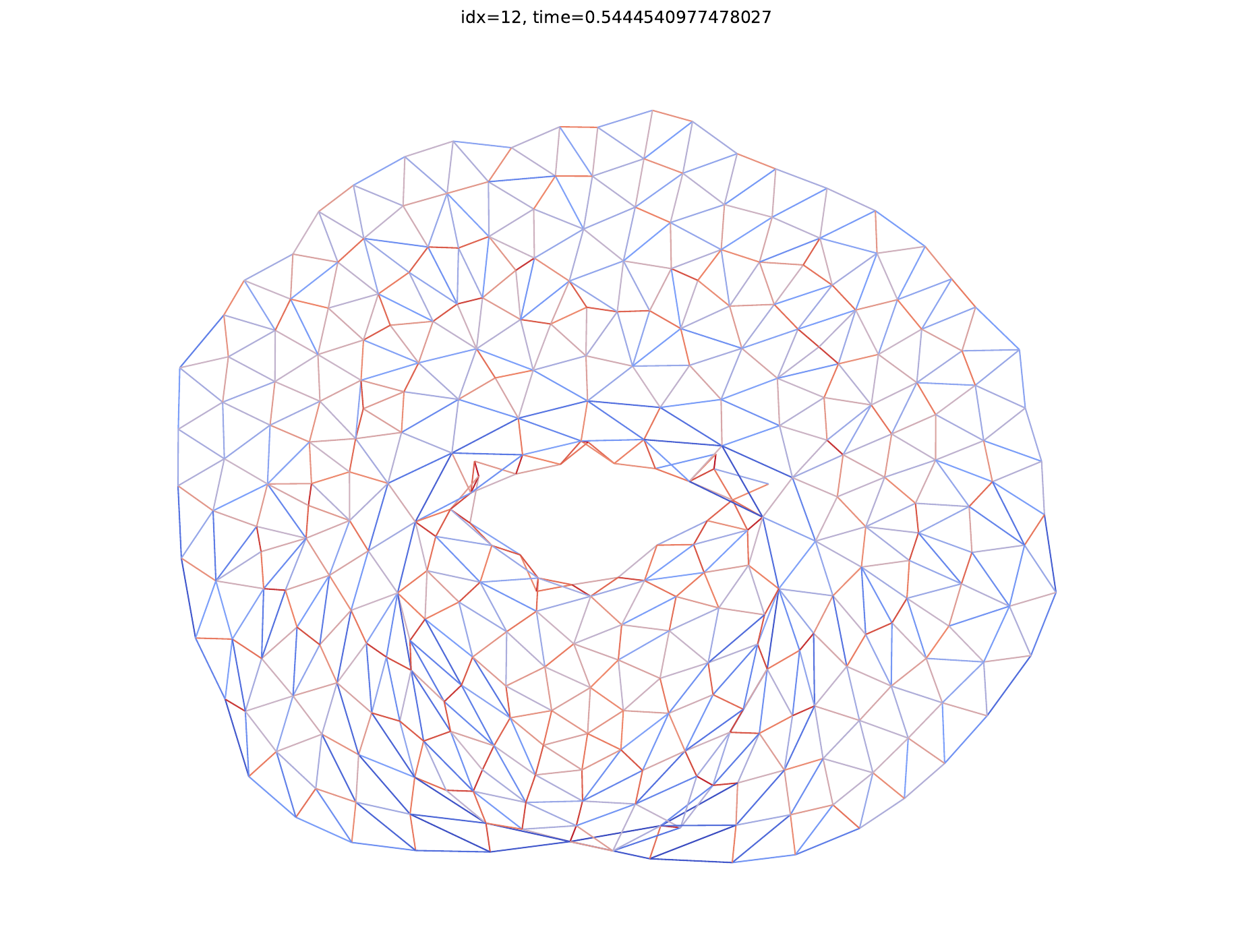} \\

&
t = 0.03s &
t = 4.55s &
t = 7.19s &
t = 0.49s &
t = 7200.00s &
t = 0.56s &
t = 0.40s &
t = 0.63s &
t = 0.36s &
t = 0.83s &
t = 0.47s &
t = 0.62s \\

\makecell{\bfseries dwt\_307\\N = 307\\M = 1108} &
\imgcell{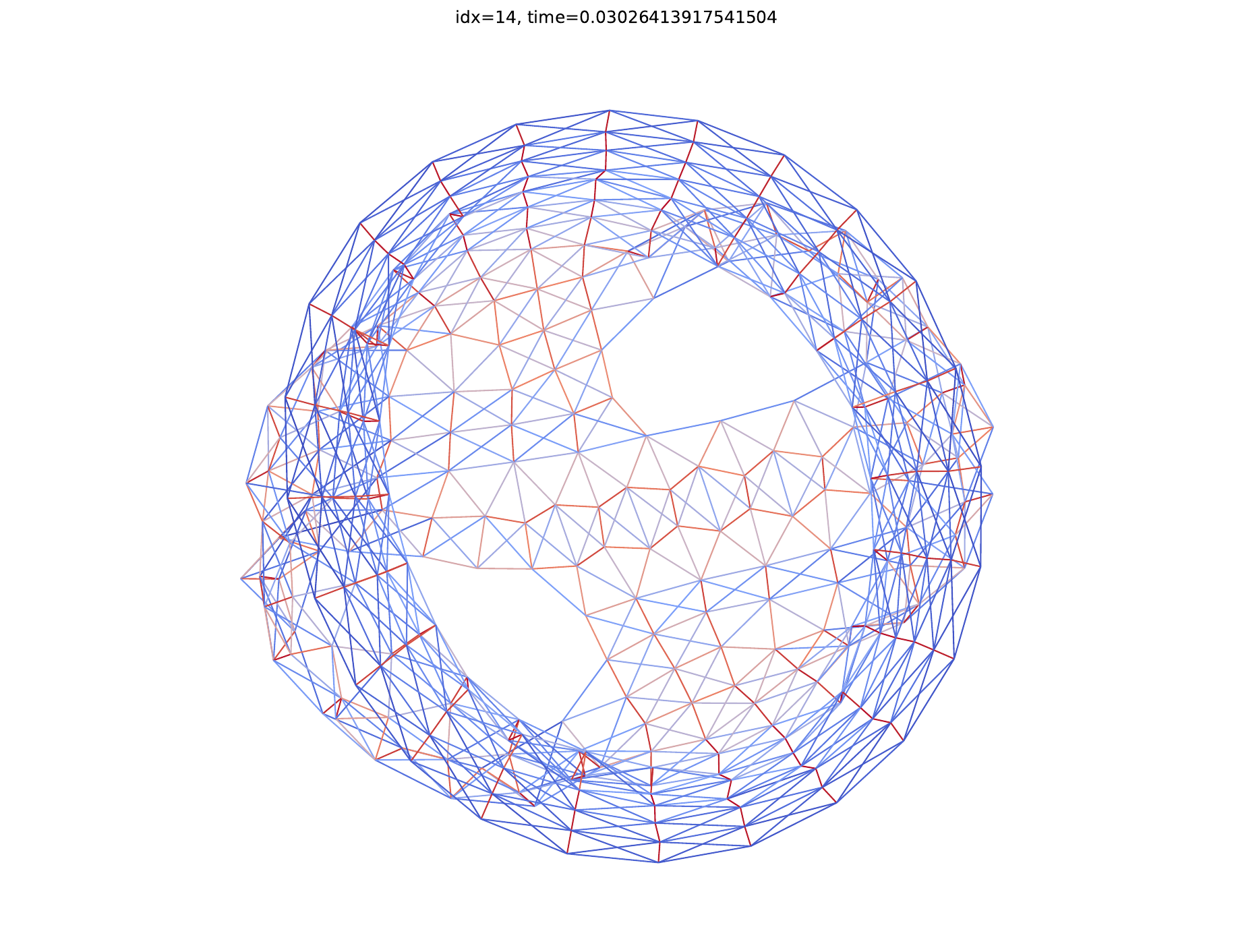} &
\imgcell{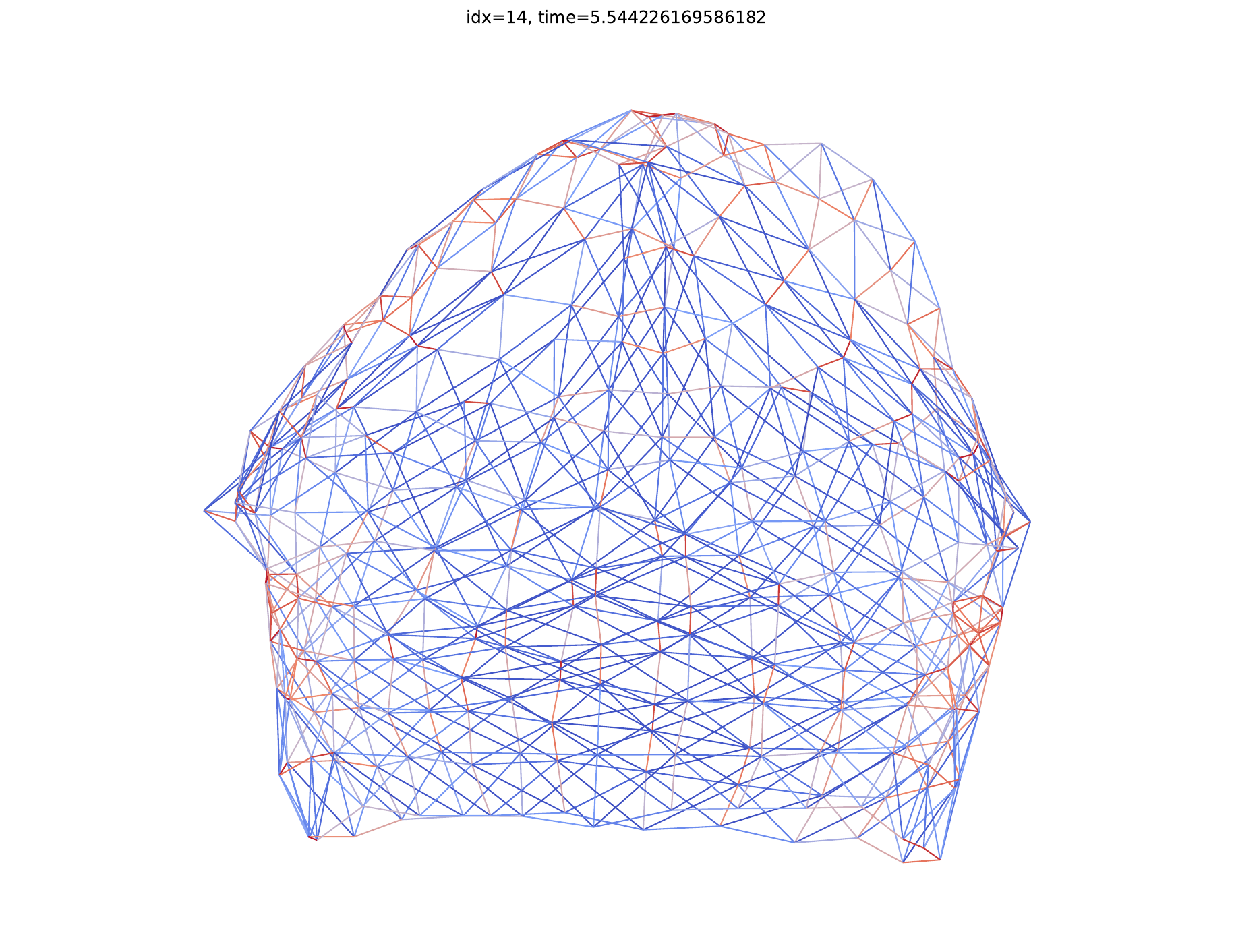} &
\imgcell{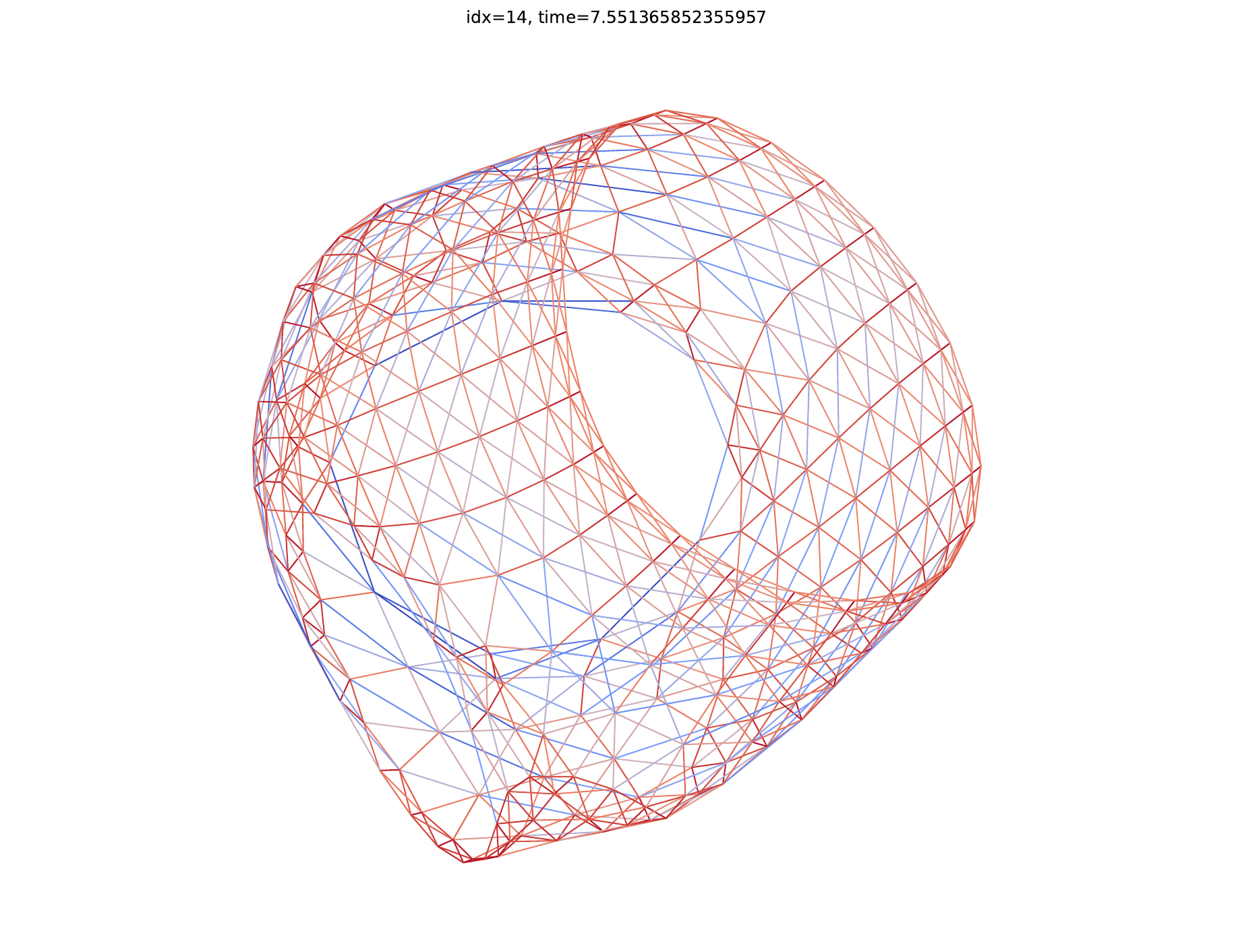} &
\imgcell{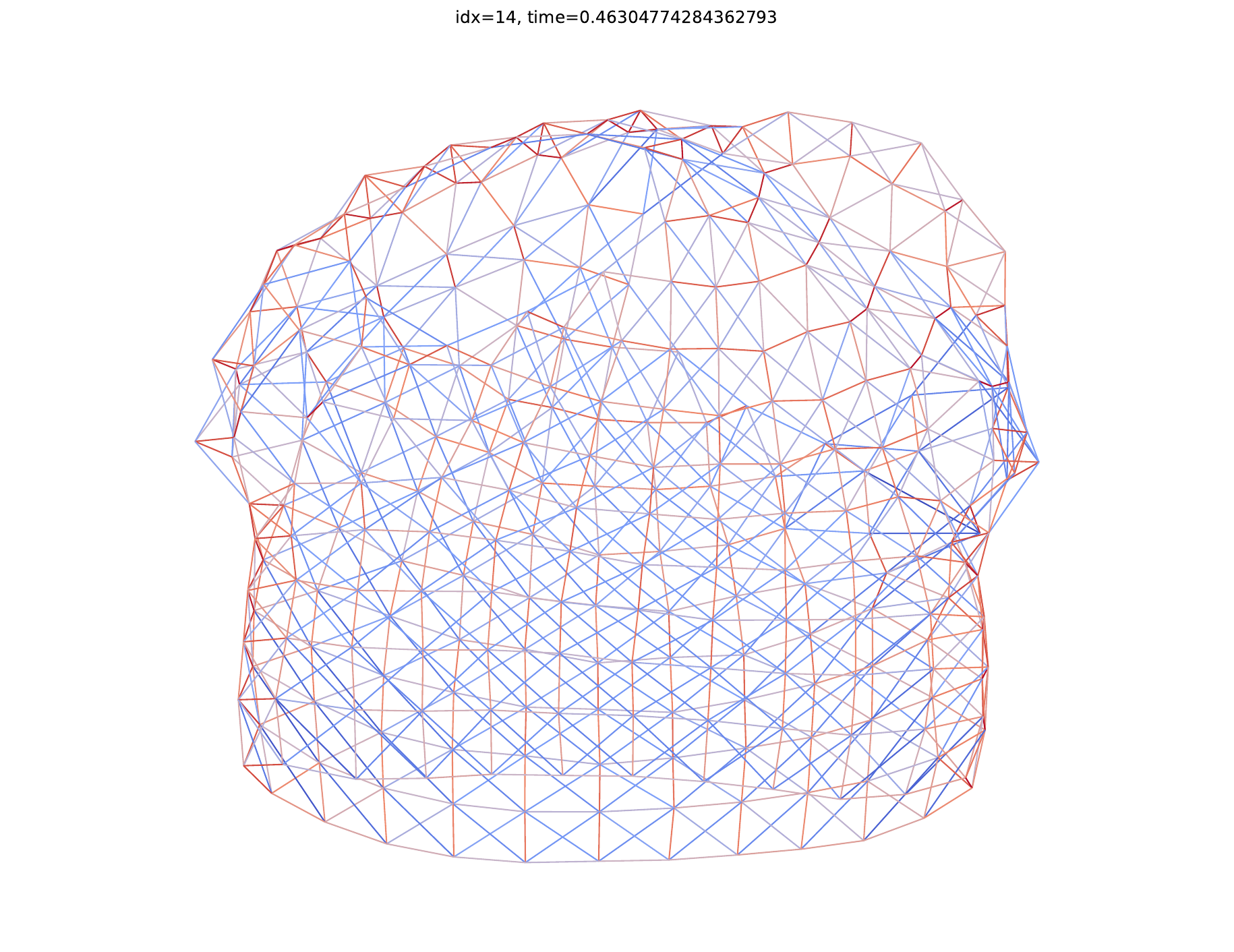} &
\imgcell{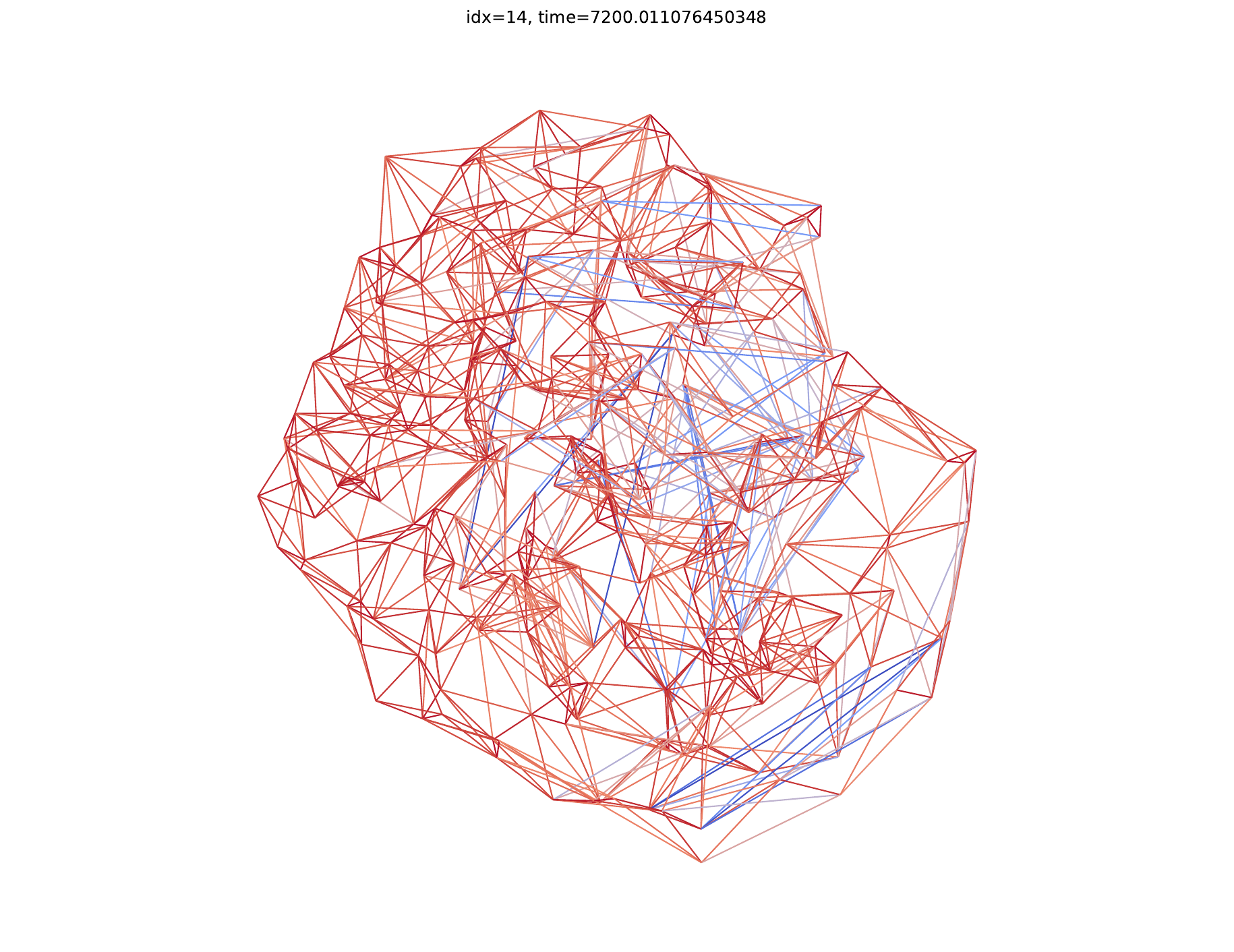} &
\imgcell{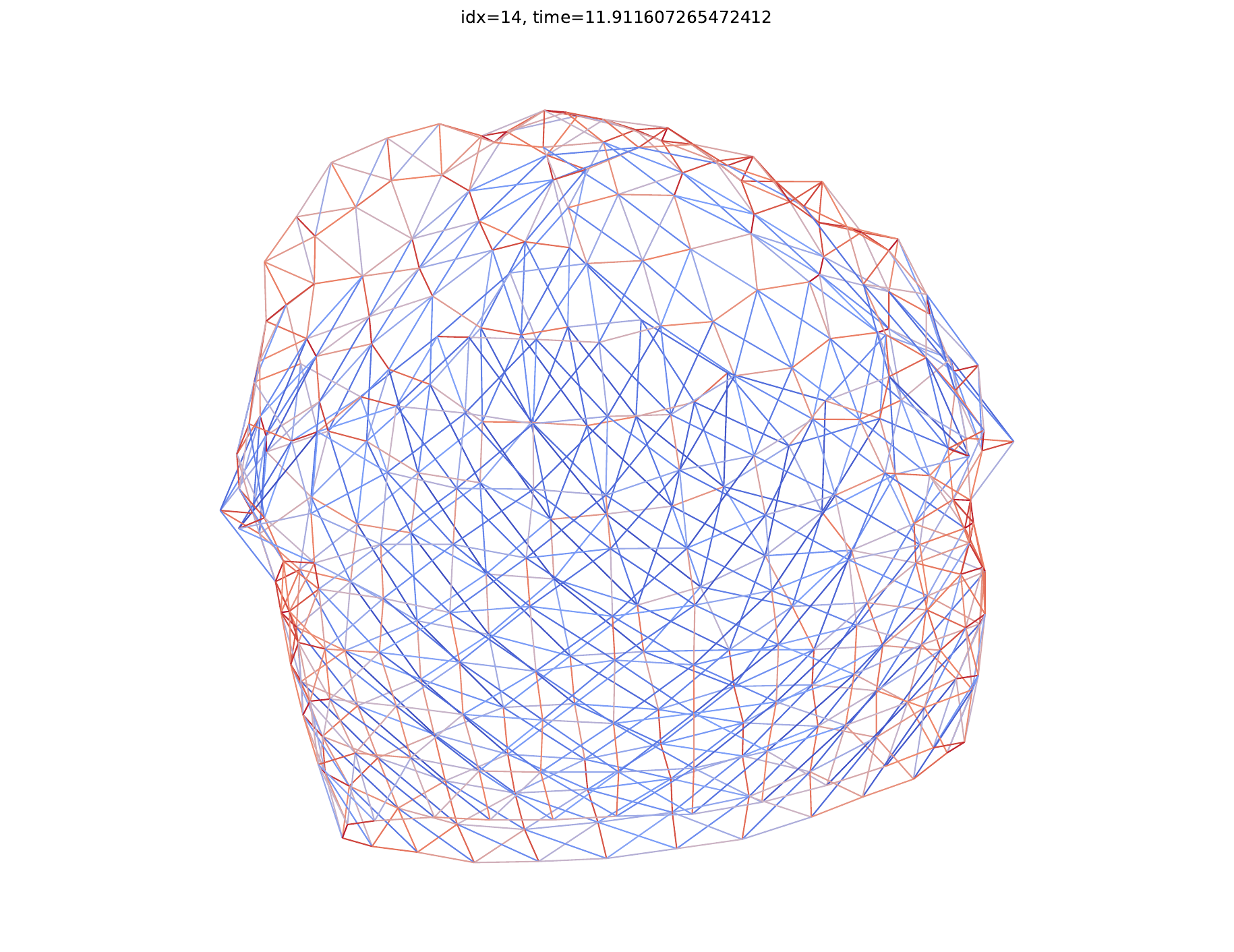} &
\imgcell{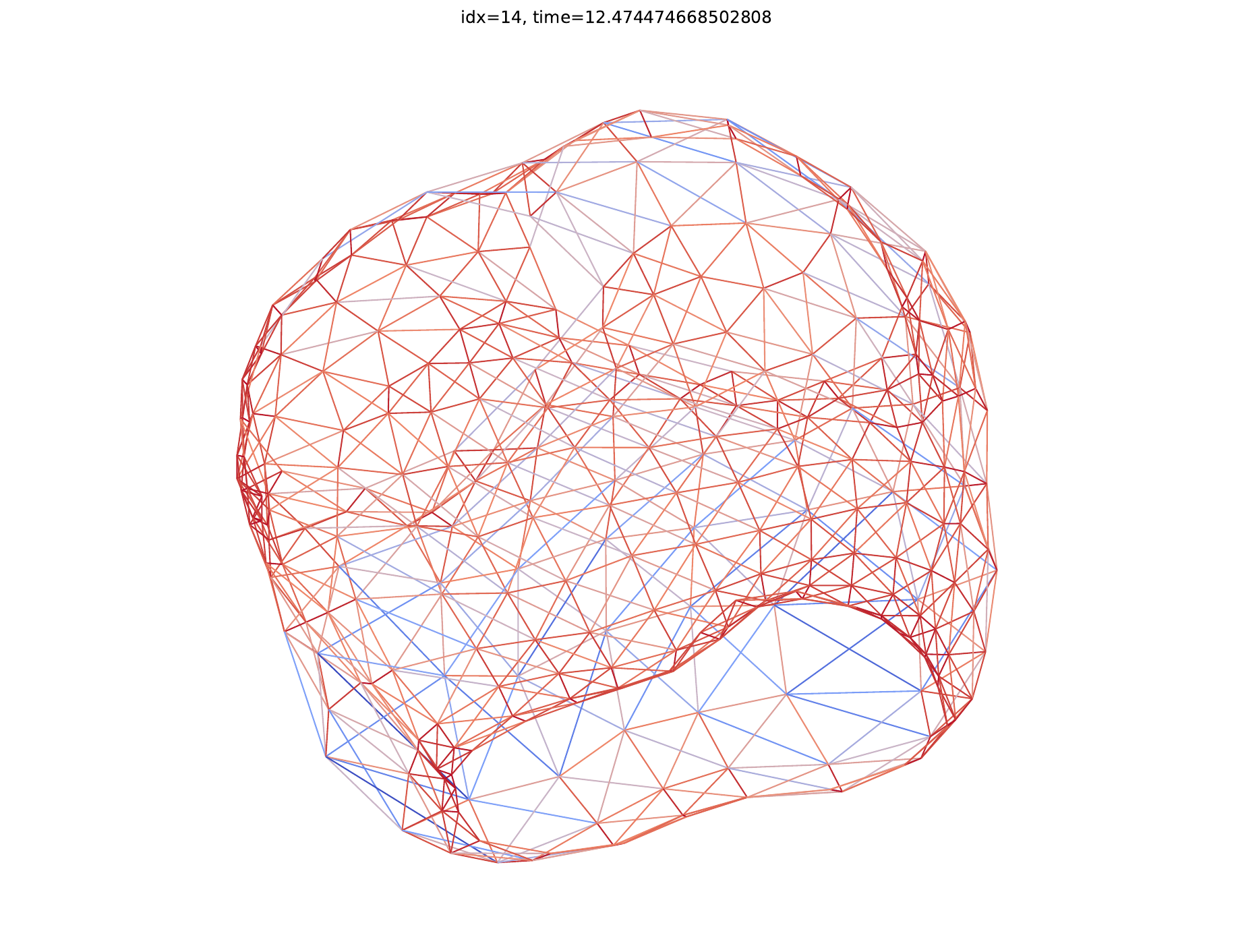} &
\imgcell{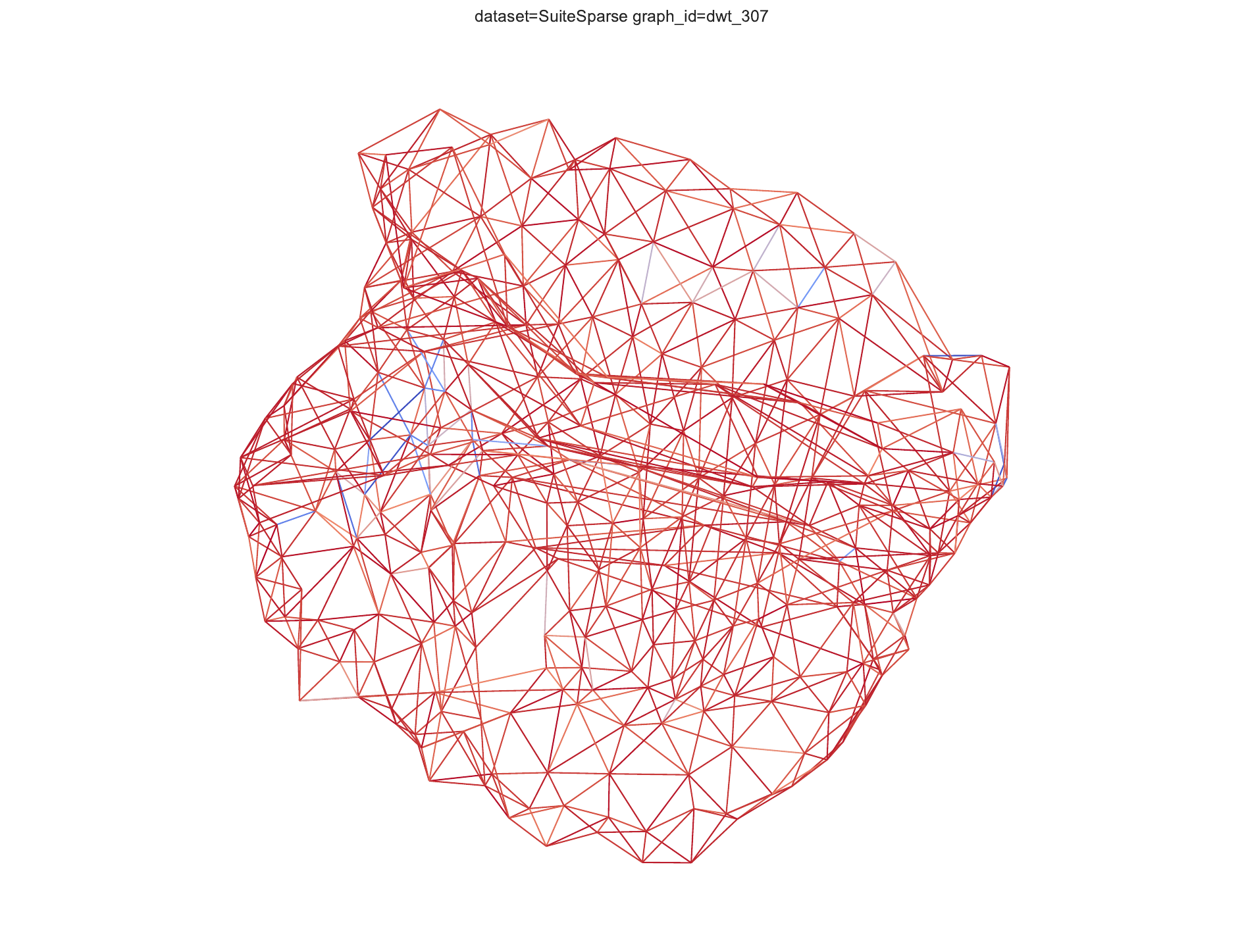} &
\imgcell{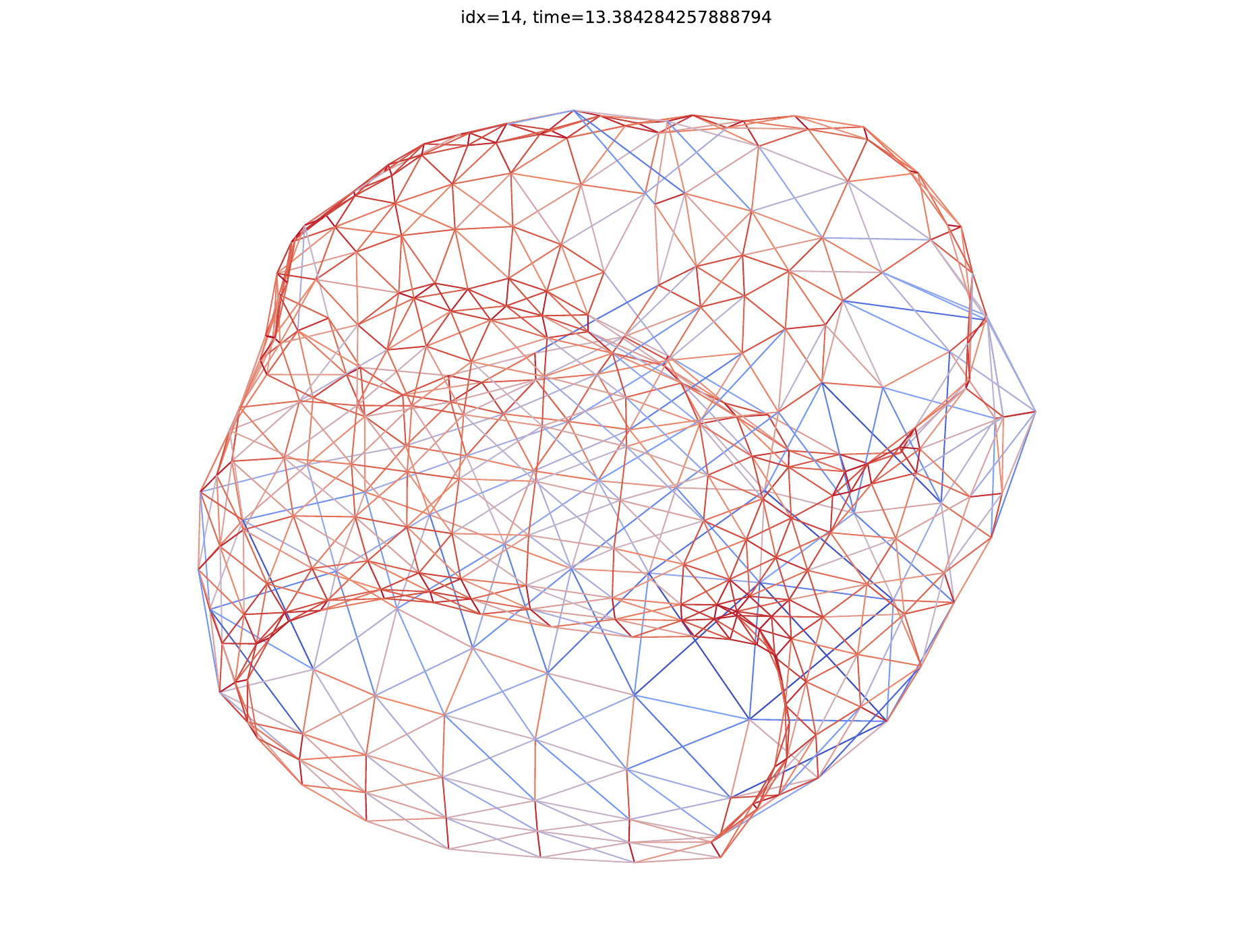} &
\imgcell{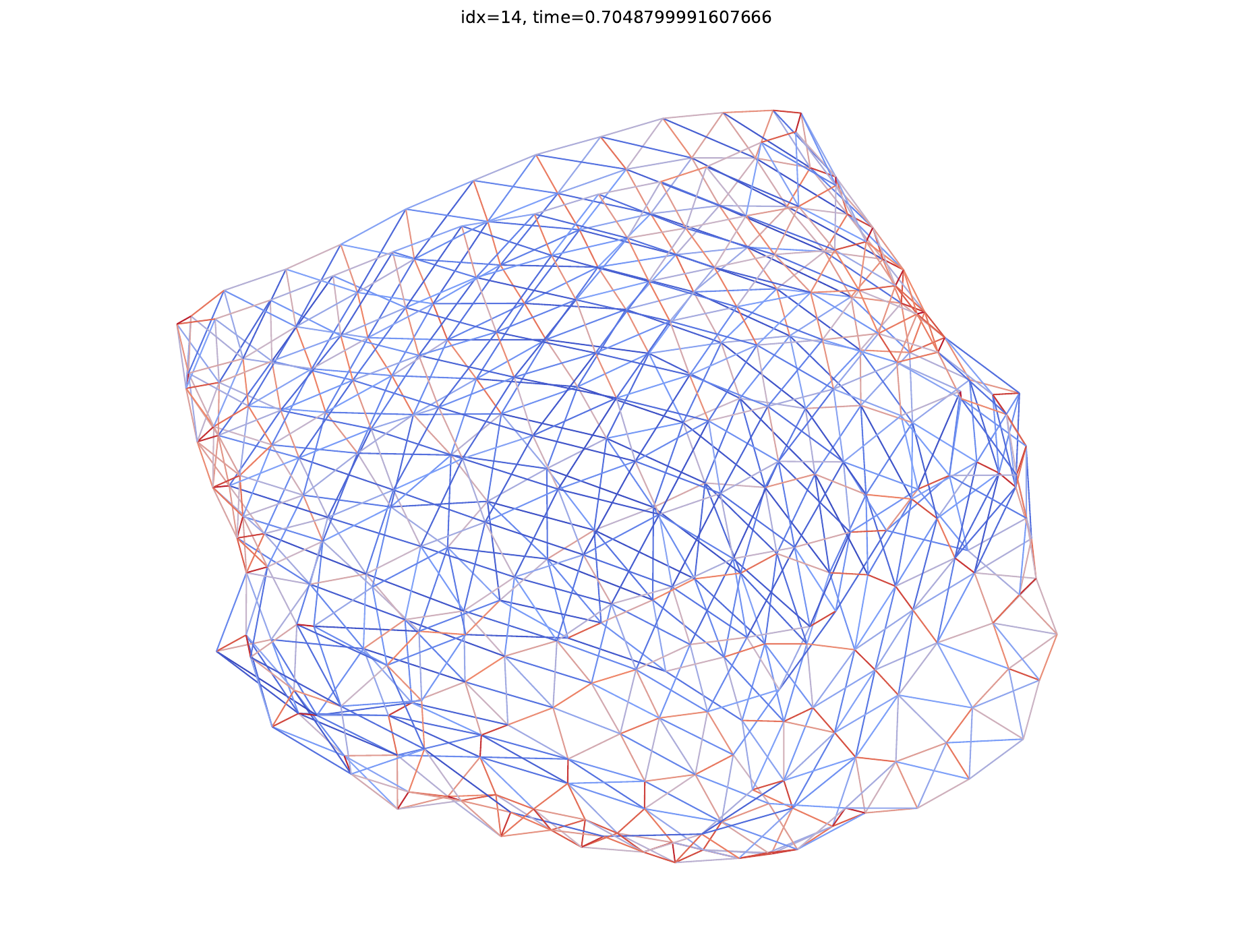} &
\imgcell{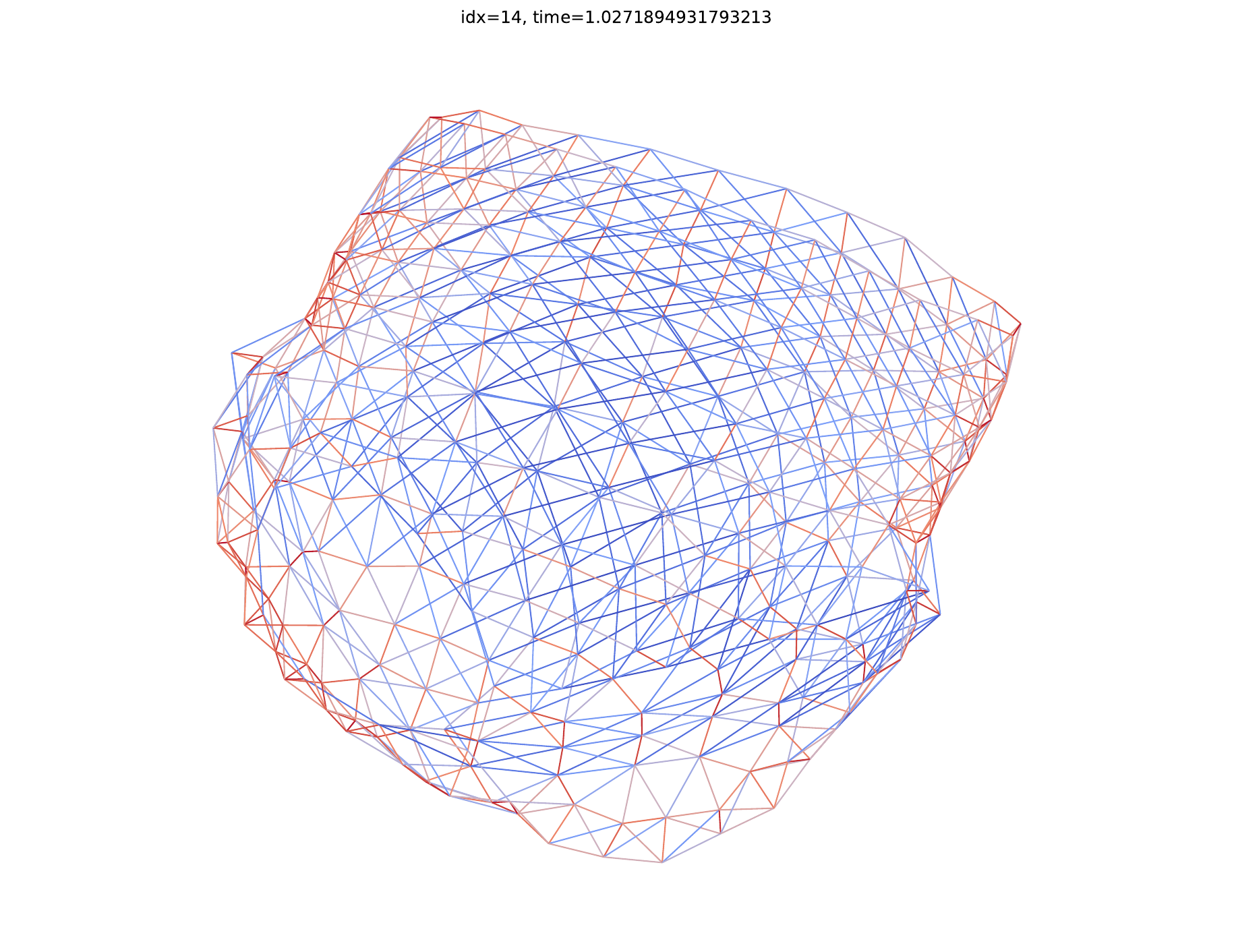} &
\imgcell{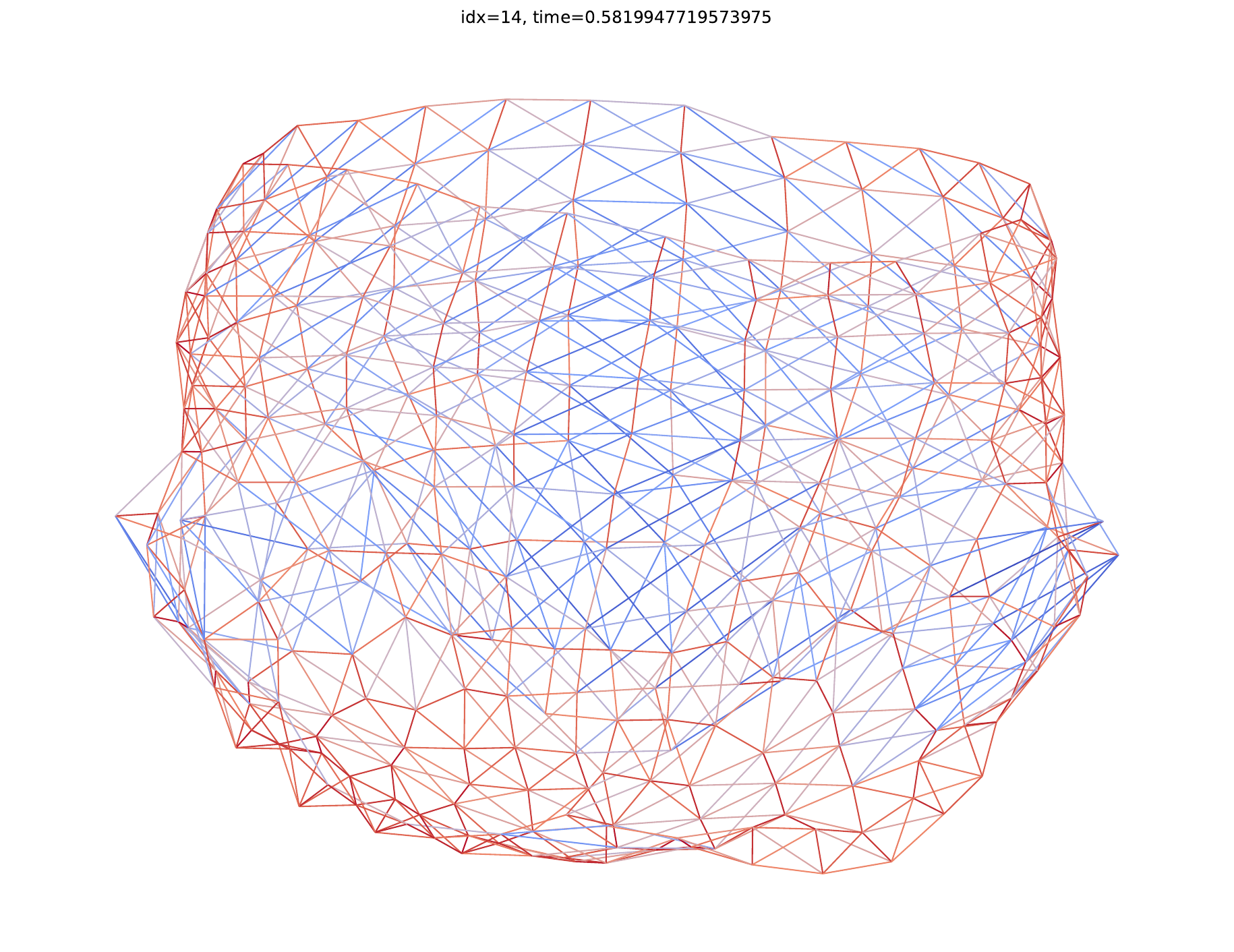} \\

&
t = 0.03s &
t = 5.54s &
t = 7.55s &
t = 0.46s &
t = 7200.00s &
t = 0.58s &
t = 0.52s &
t = 0.55s &
t = 0.30s &
t = 0.60s &
t = 0.51s &
t = 0.64s \\

\makecell{\bfseries cavity04\\N = 317\\M = 4237} &
\imgcell{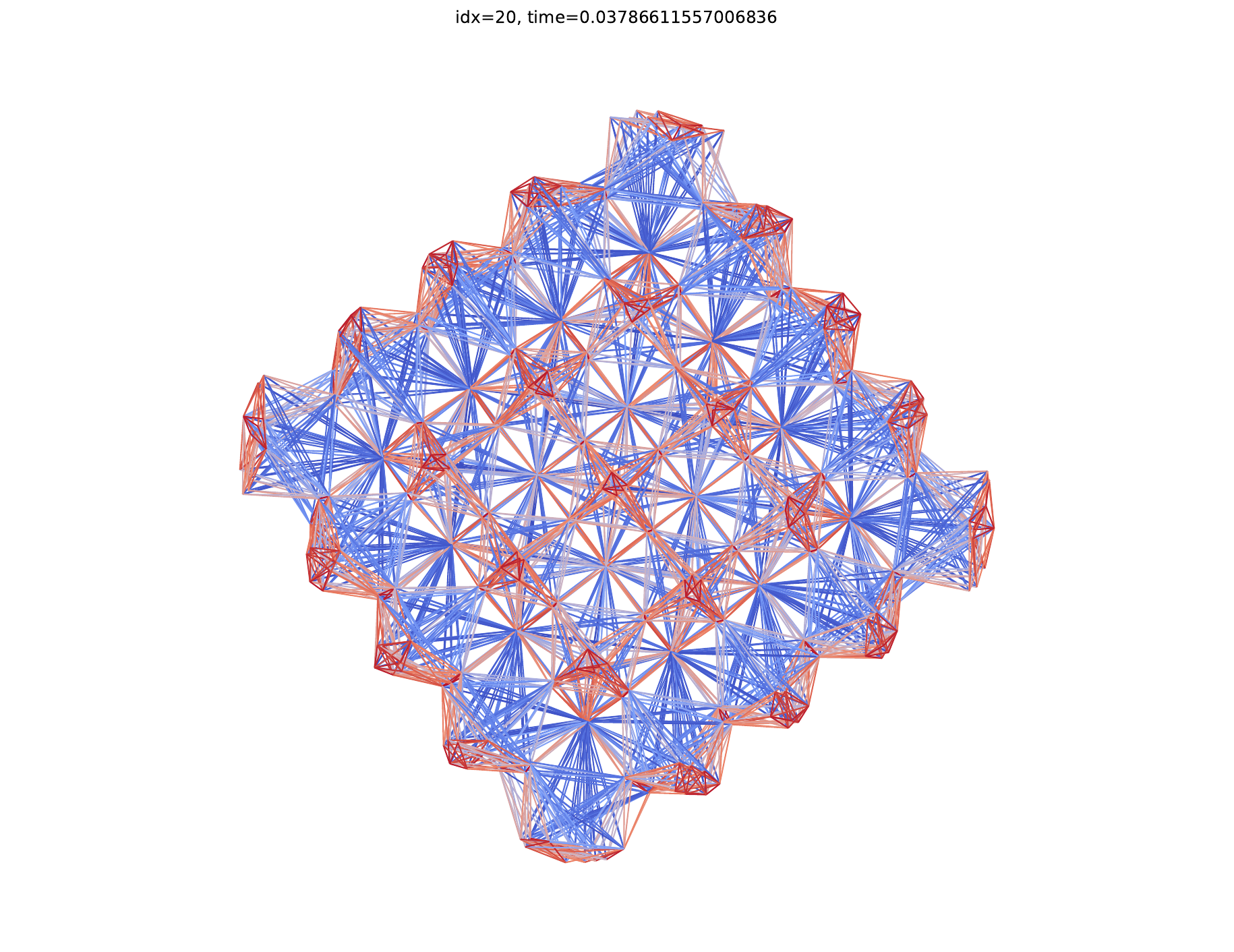} &
\imgcell{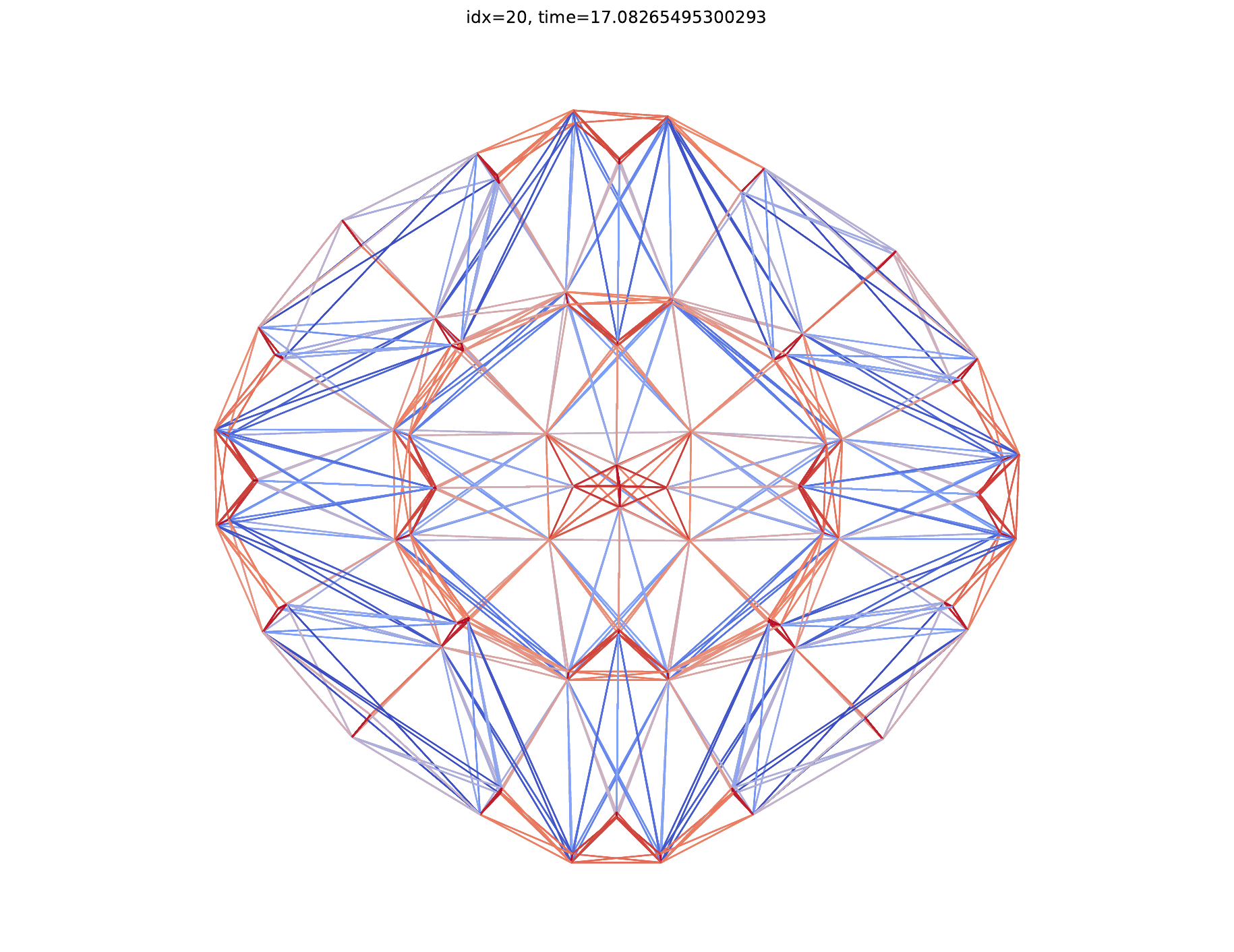} &
\imgcell{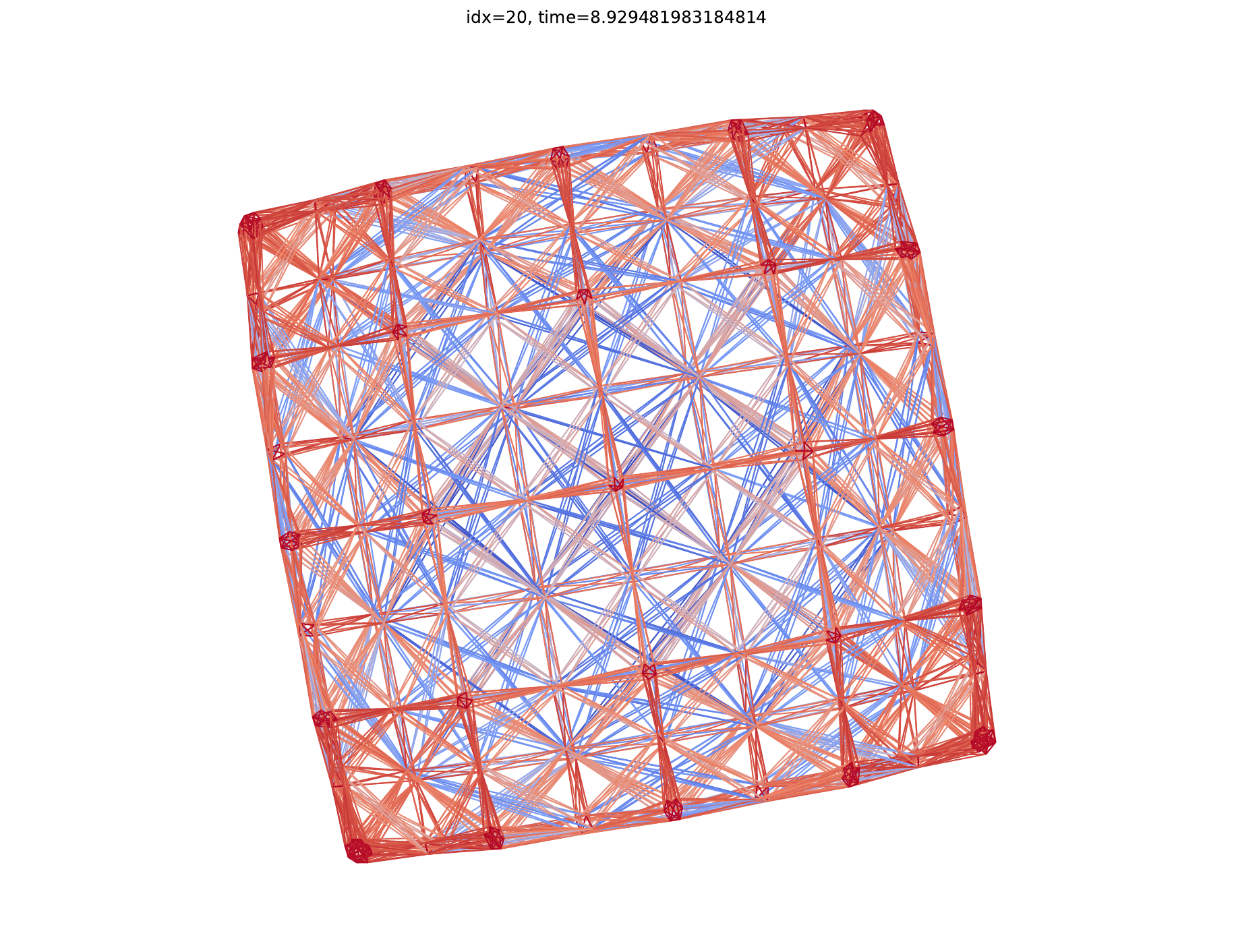} &
\imgcell{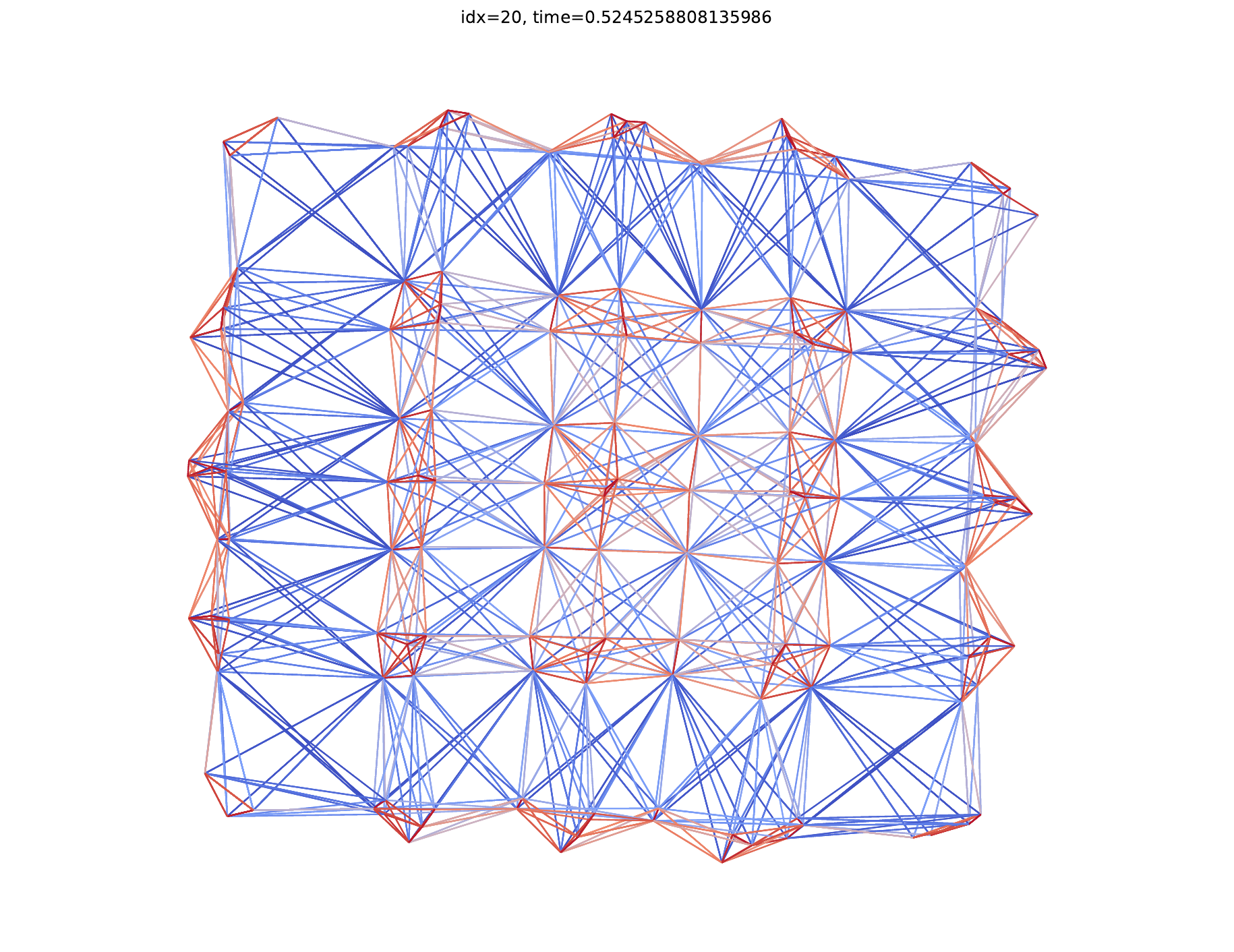} &
\imgcell{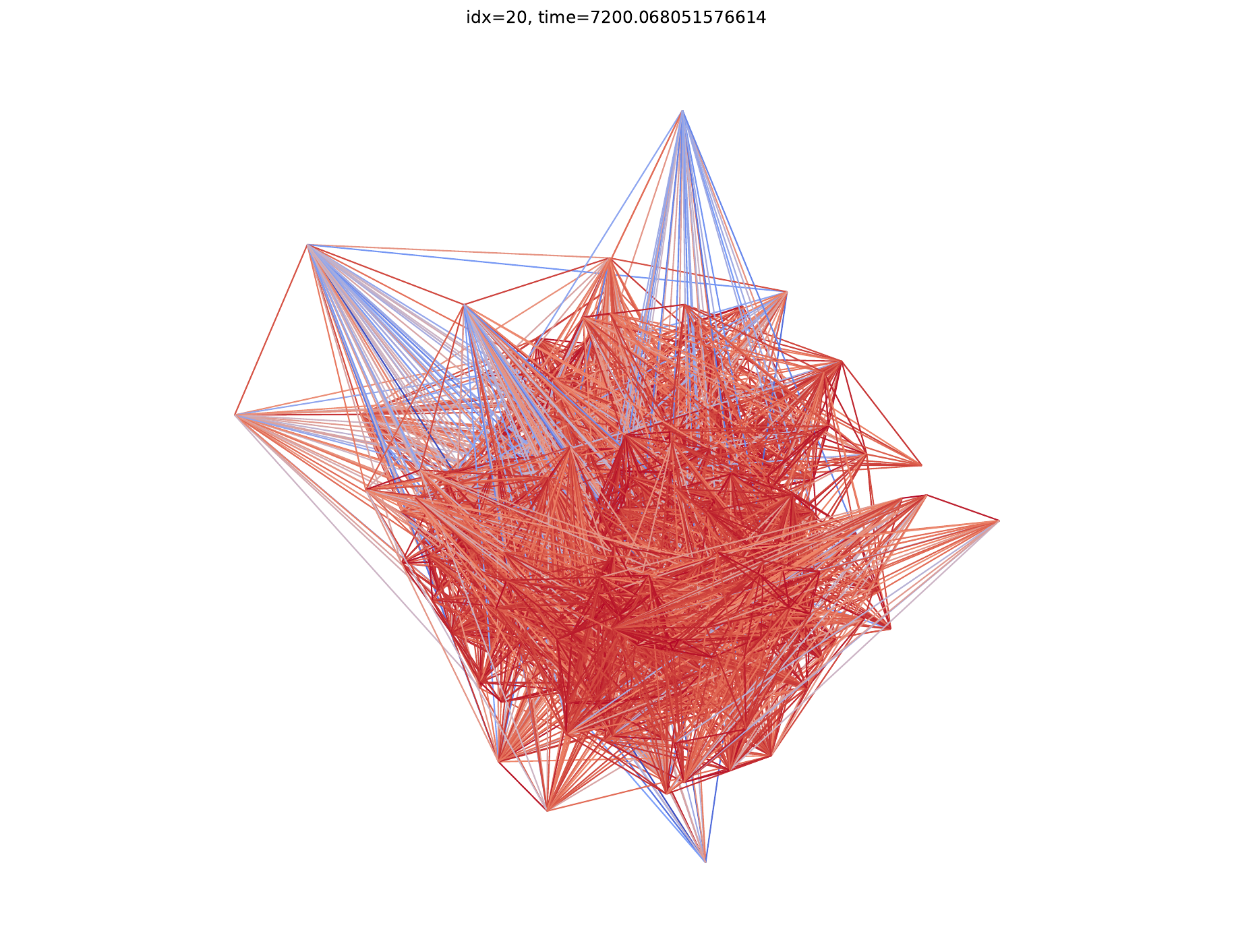} &
\imgcell{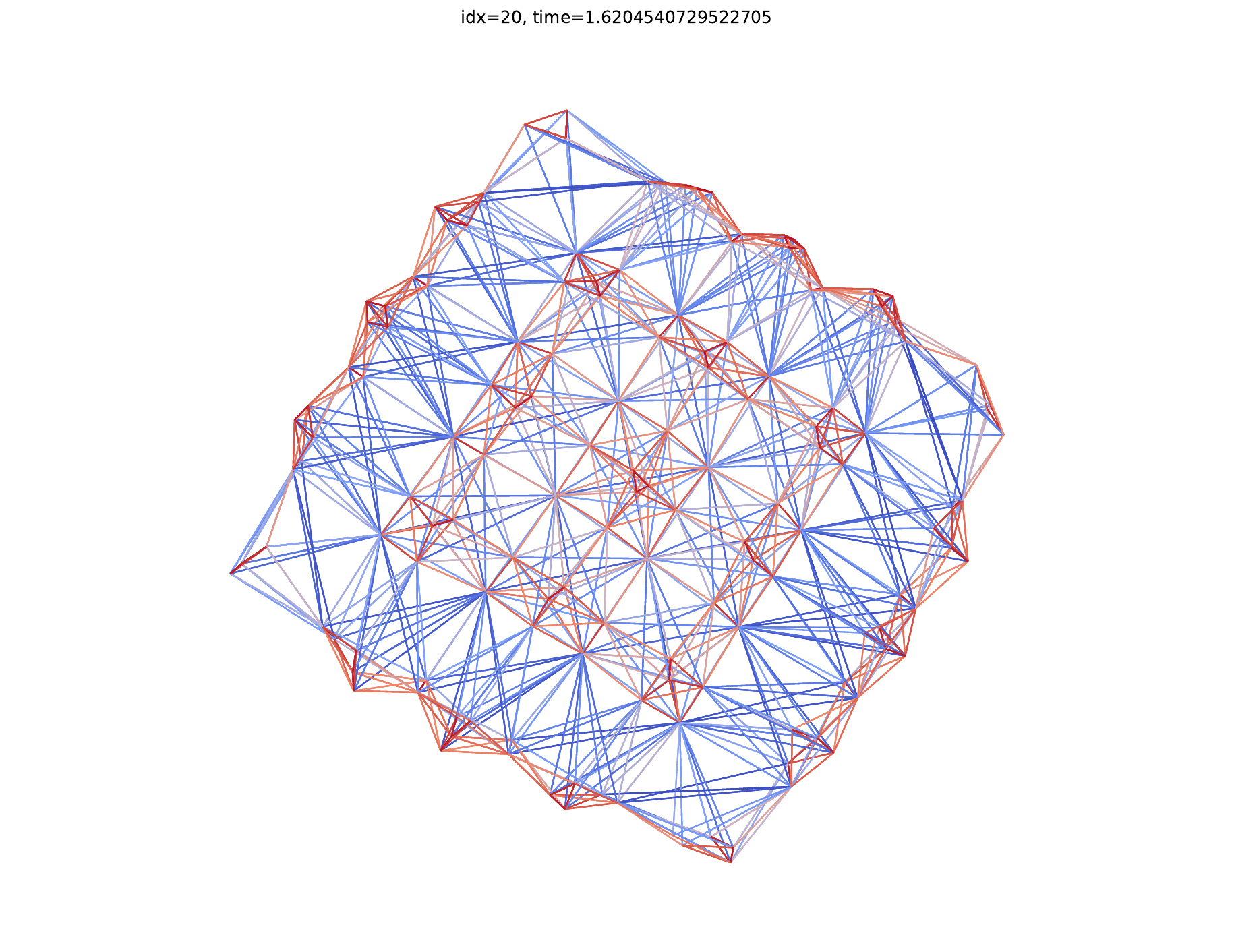} &
\imgcell{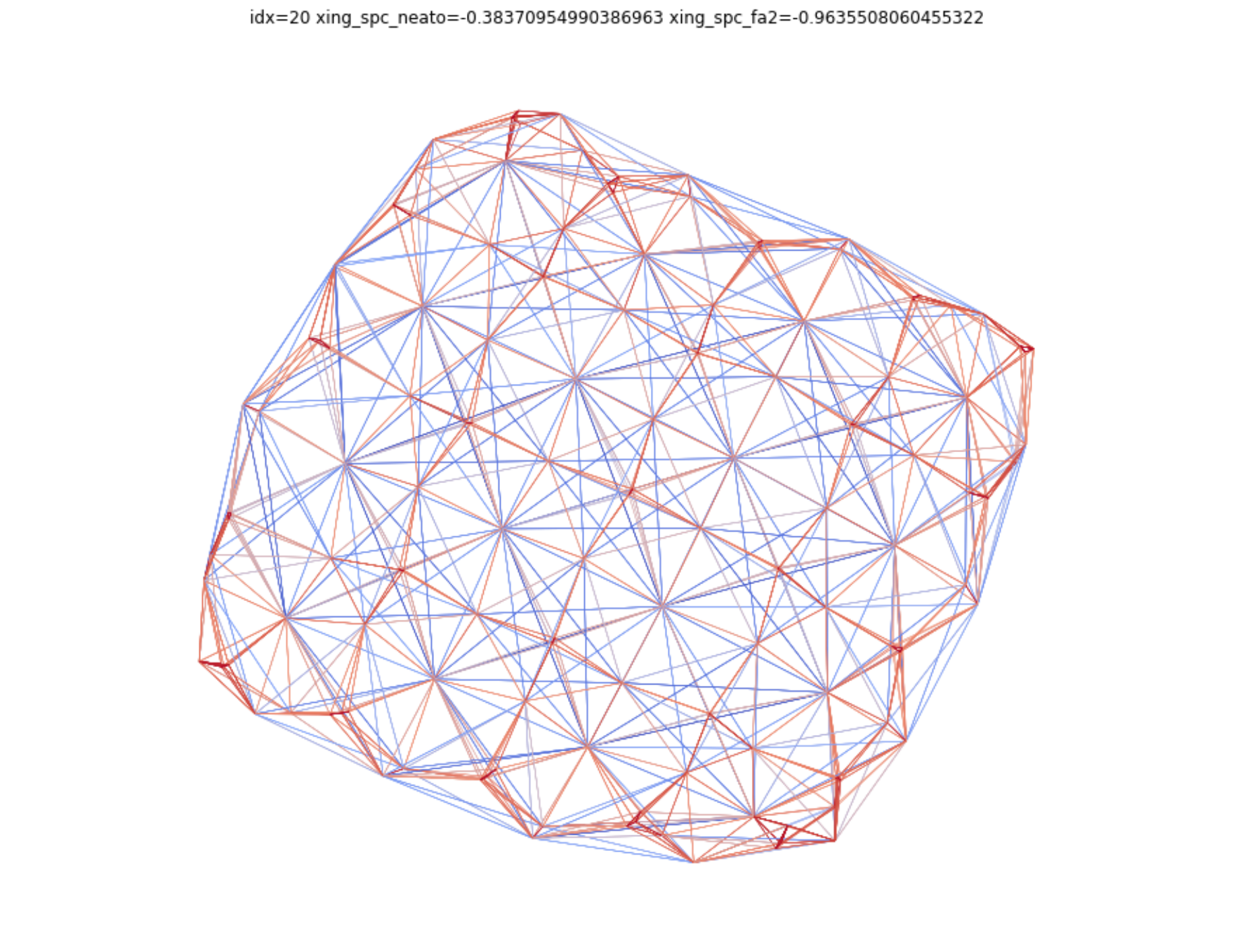} &
\imgcell{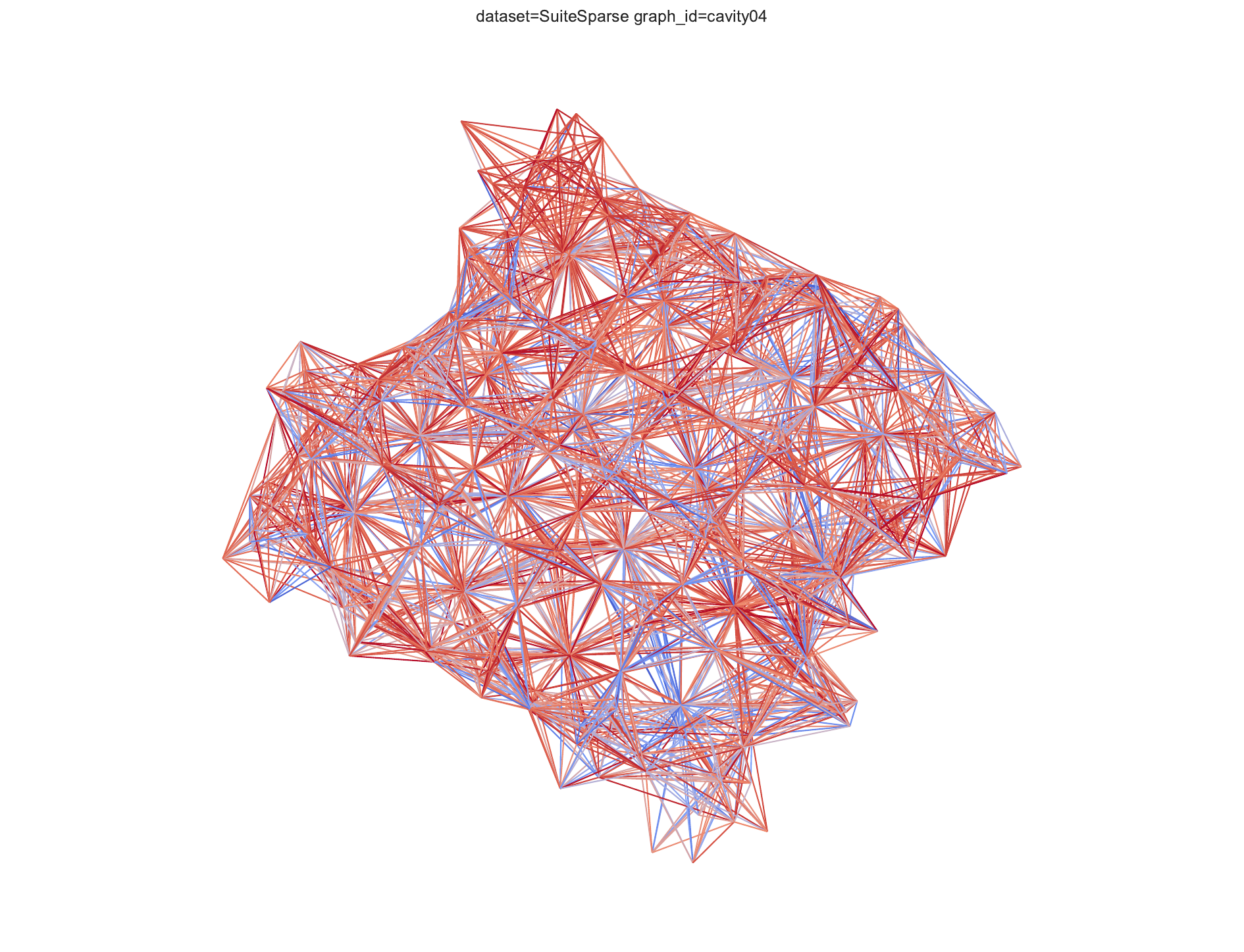} &
\imgcell{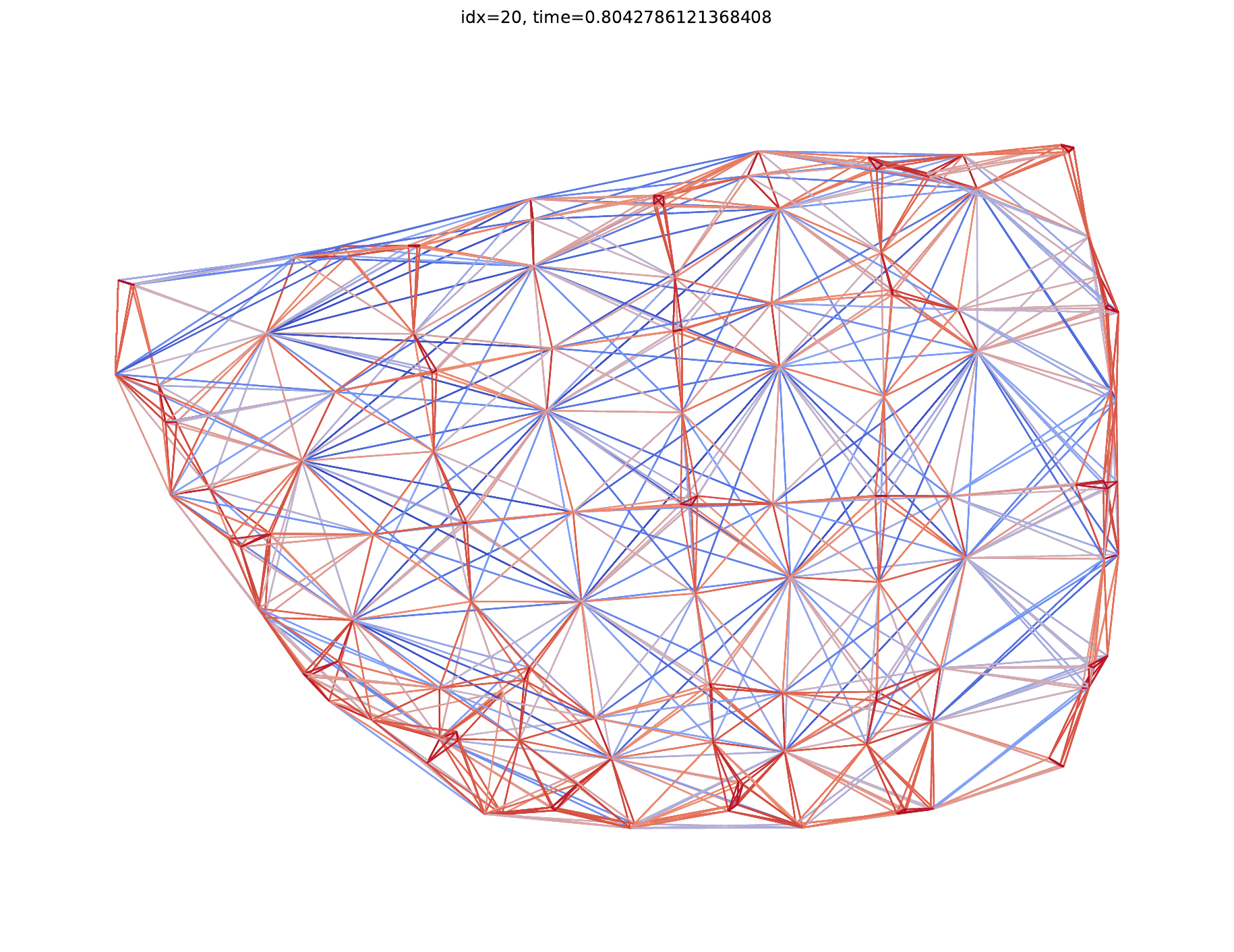} &
\imgcell{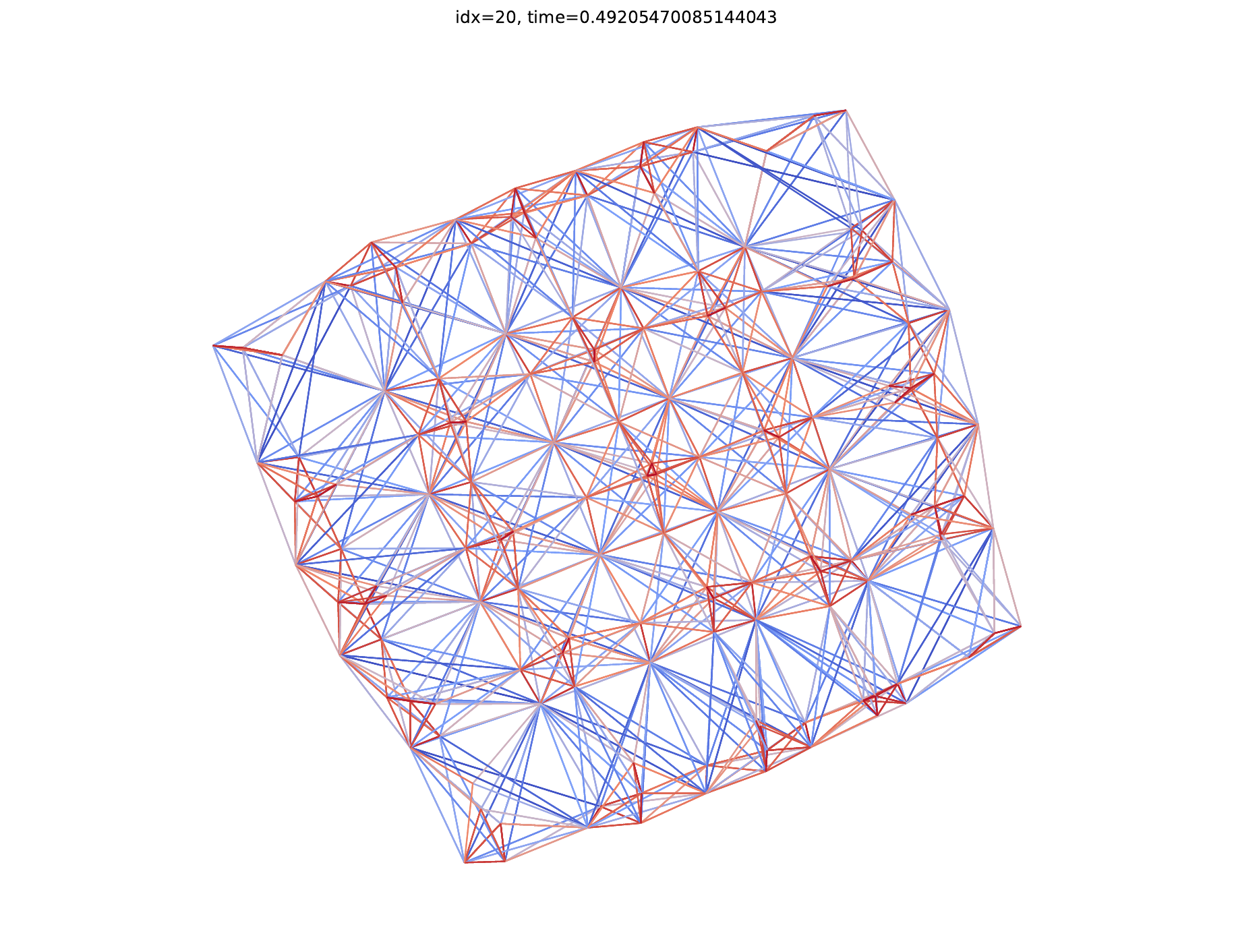} &
\imgcell{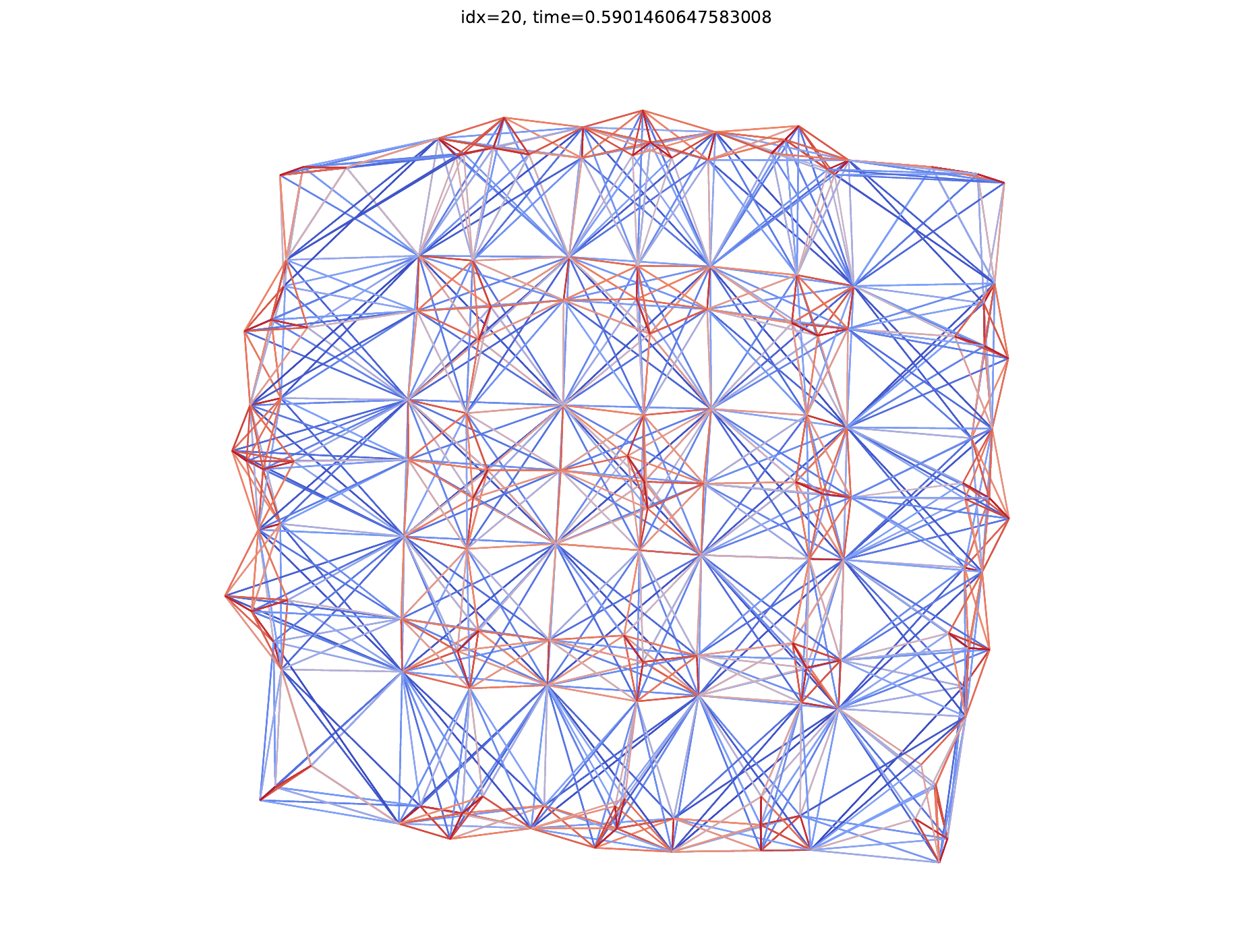} &
\imgcell{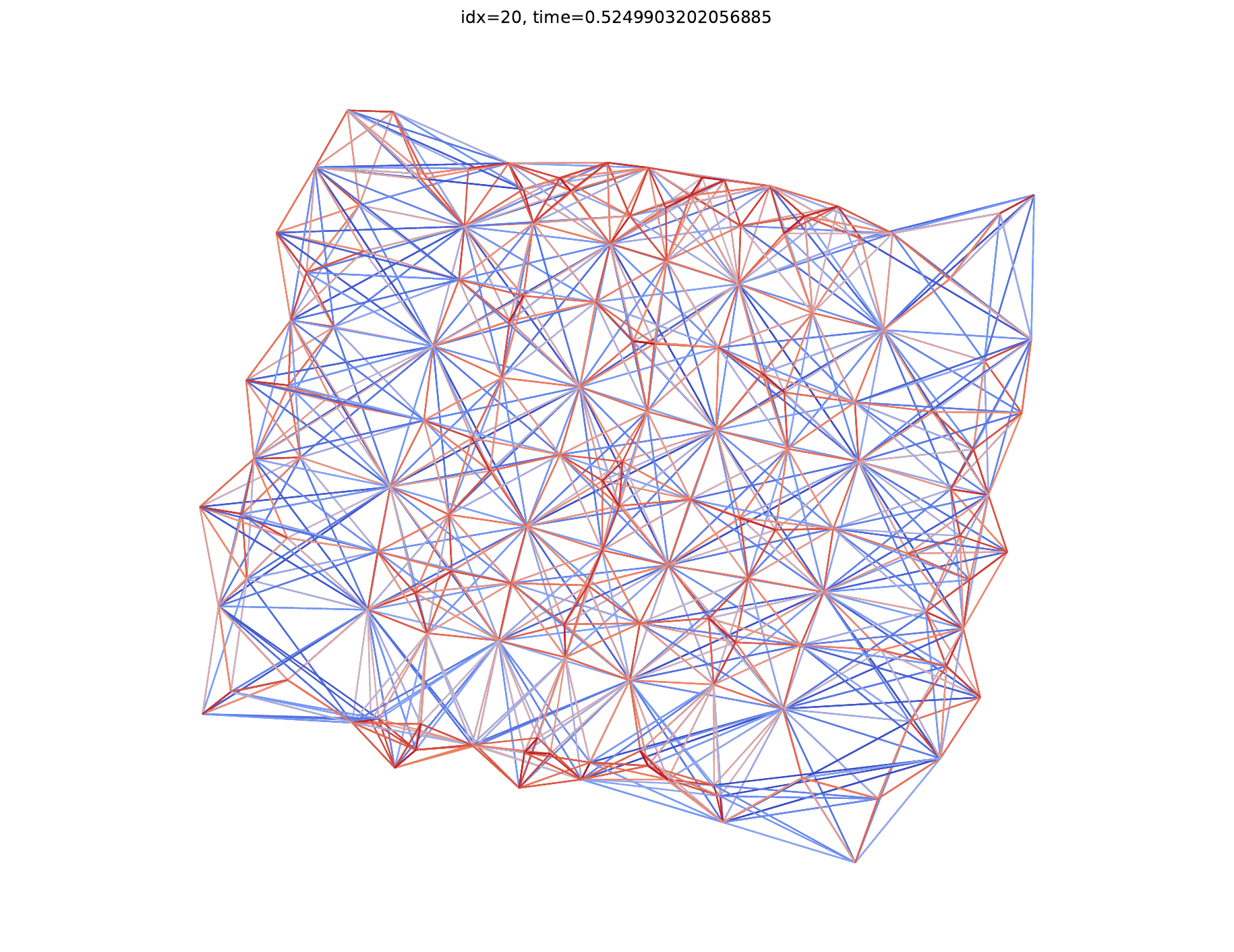} \\

&
t = 0.04s &
t = 17.08s &
t = 8.93s &
t = 0.52s &
t = 7200.00s &
t = 0.42s &
t = 0.60s &
t = 0.54s &
t = 0.62s &
t = 0.70s &
t = 0.49s &
t = 0.52s \\

\makecell{\bfseries qc324\\N = 324\\M = 13203} &
\imgcell{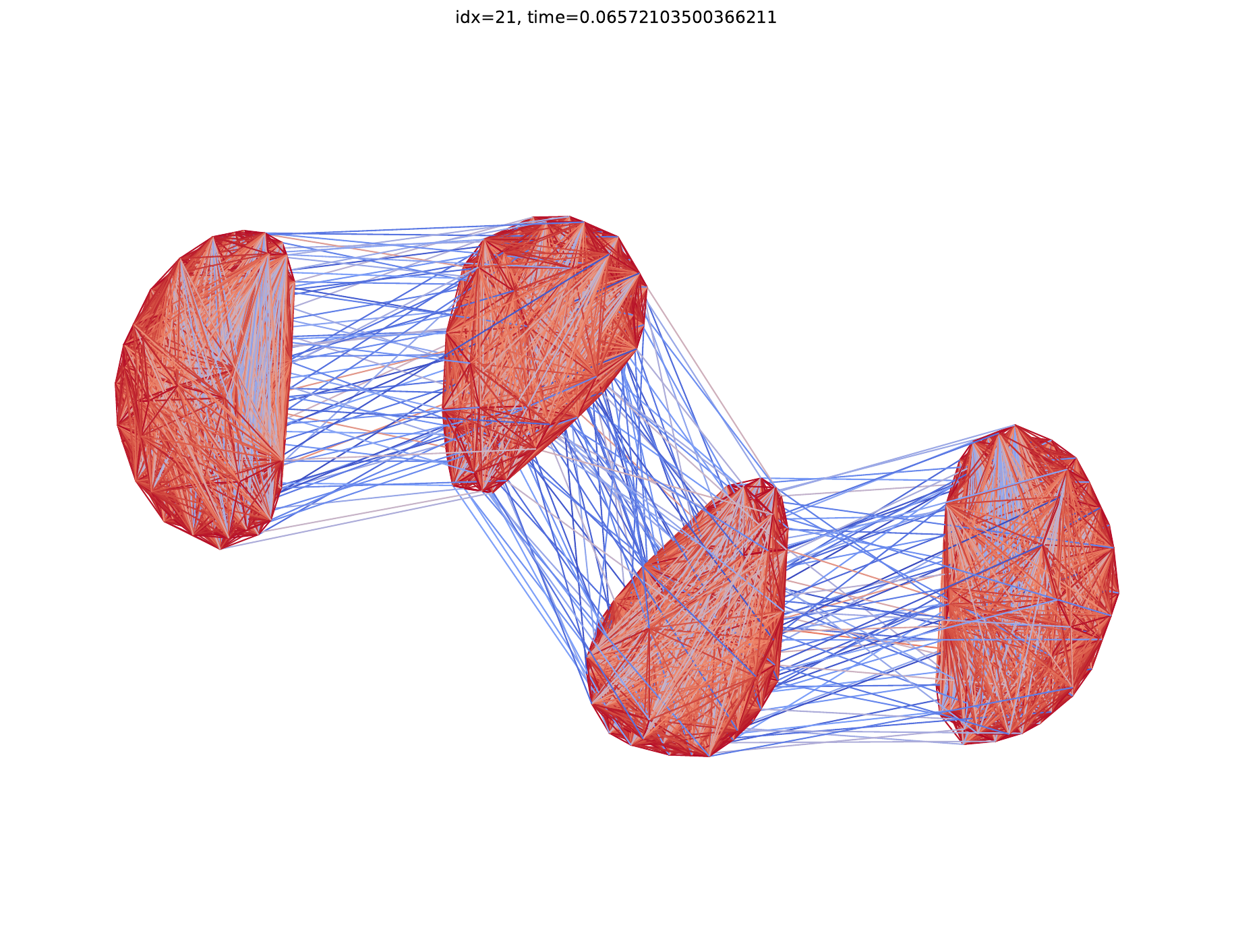} &
\imgcell{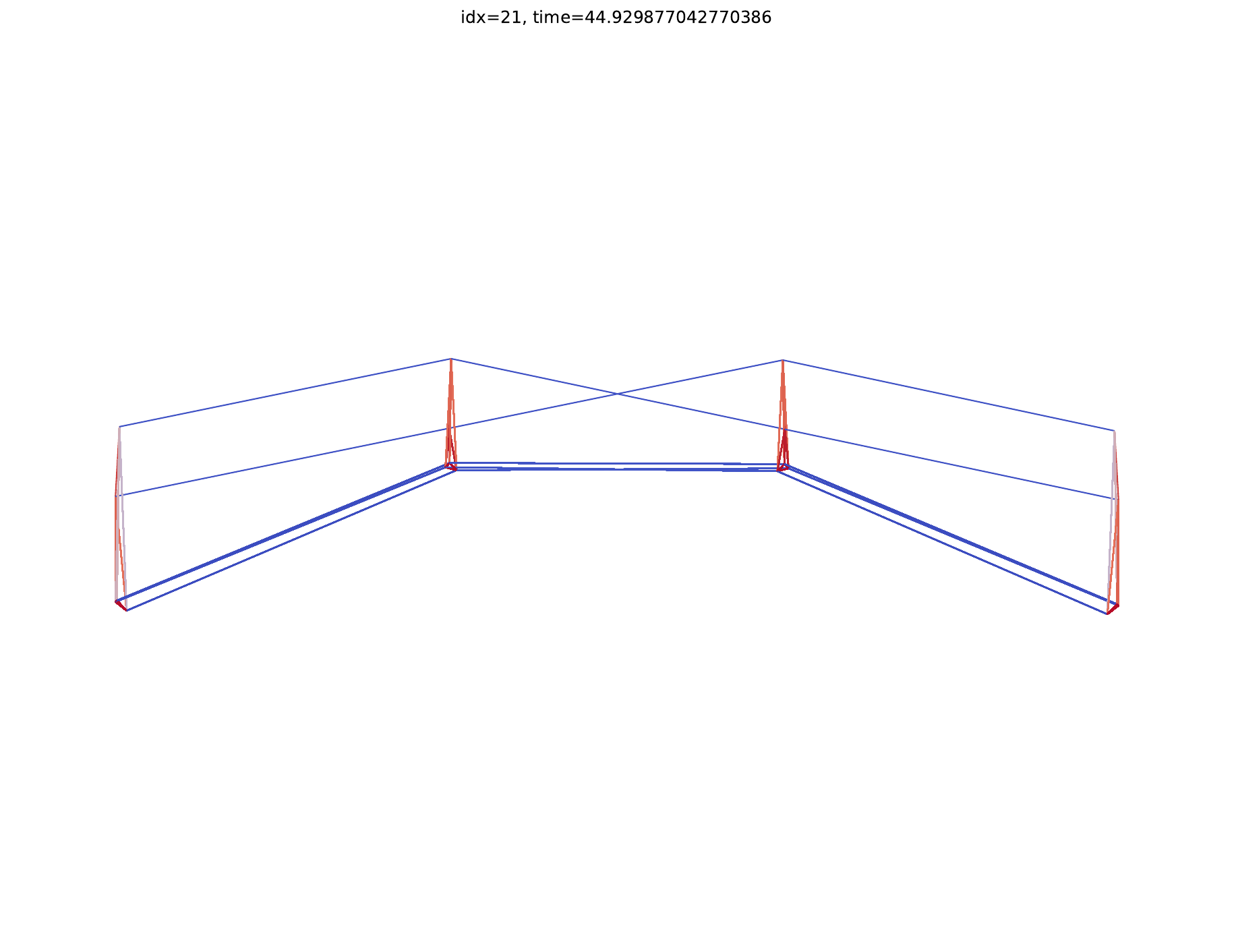} &
\imgcell{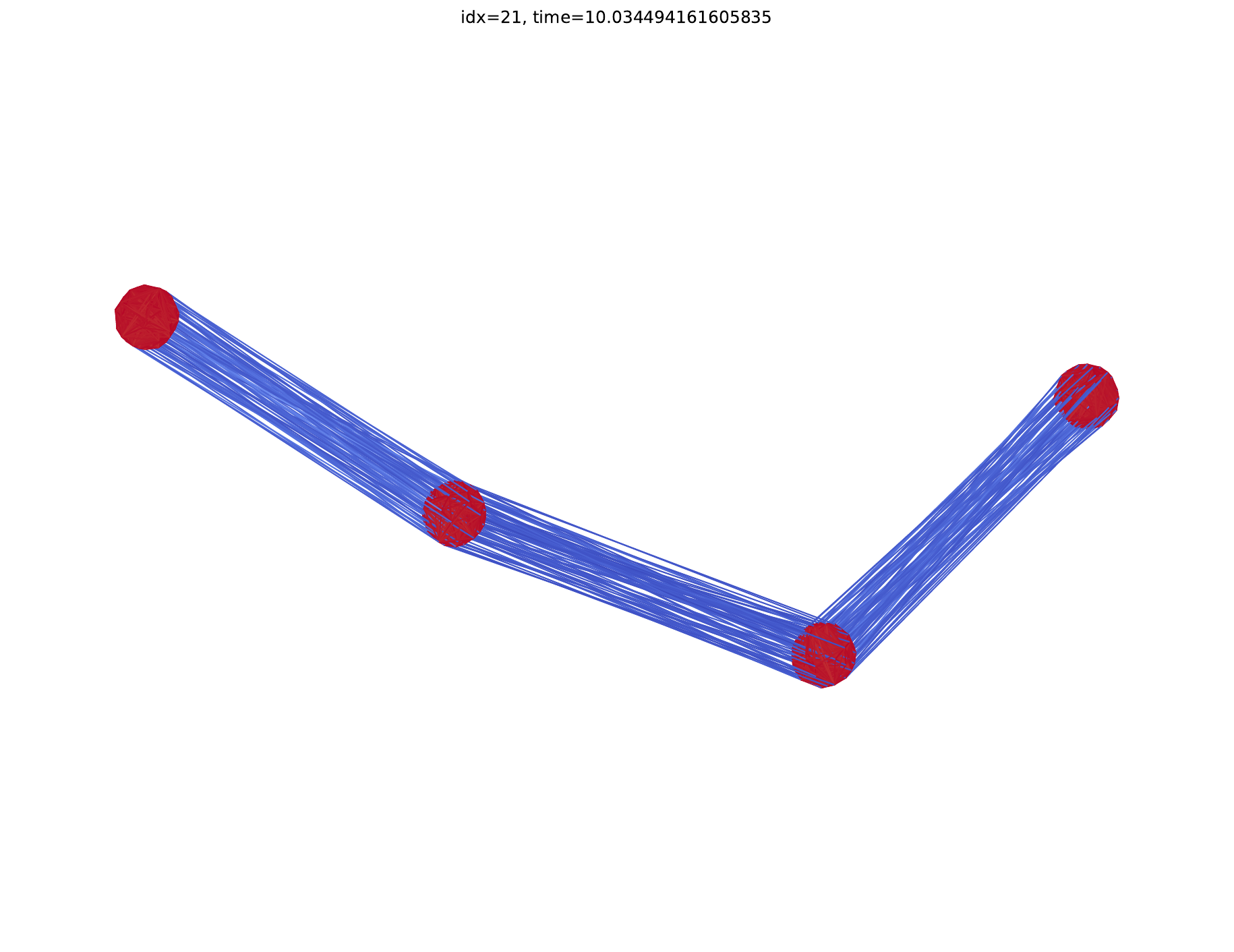} &
\imgcell{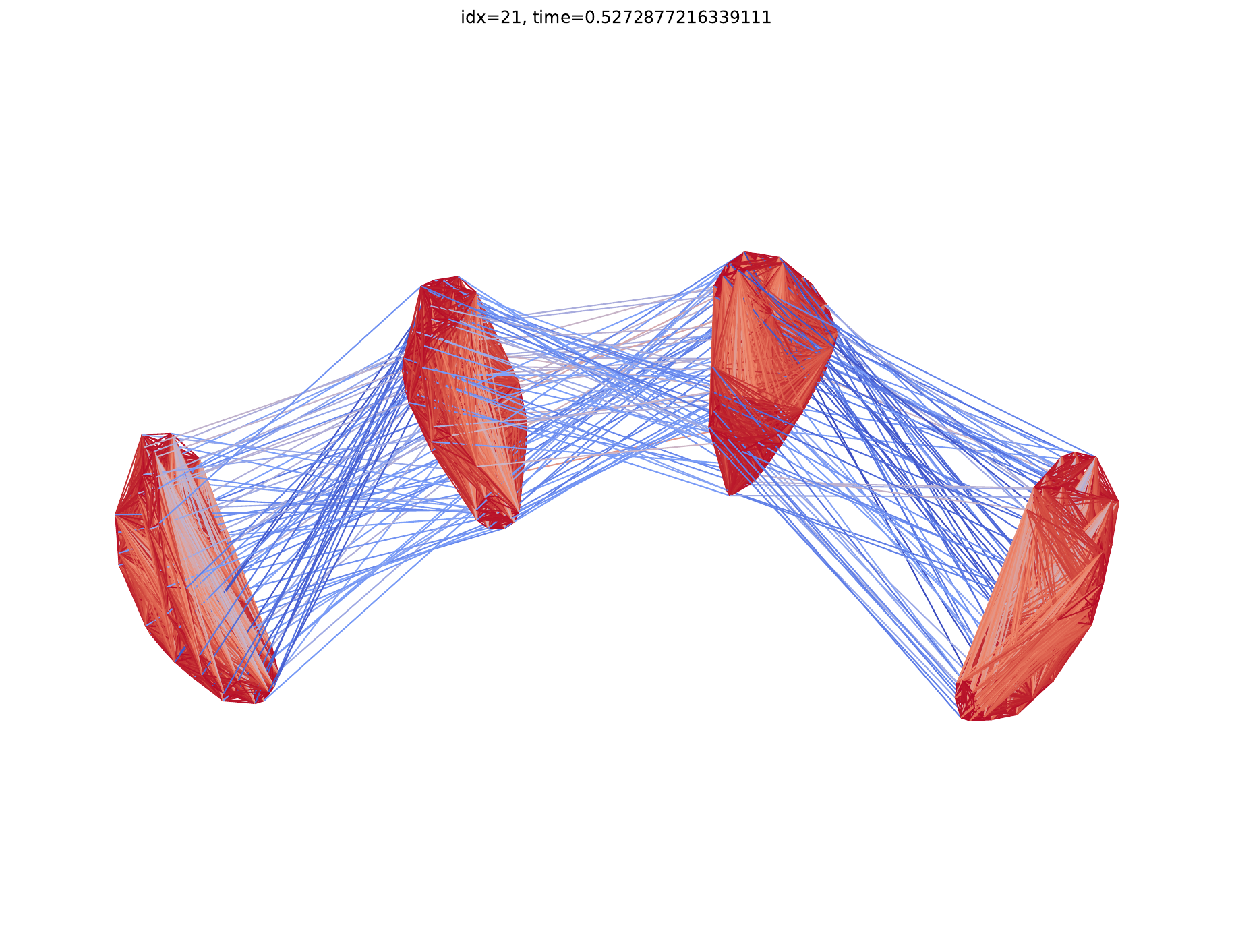} &
\imgcell{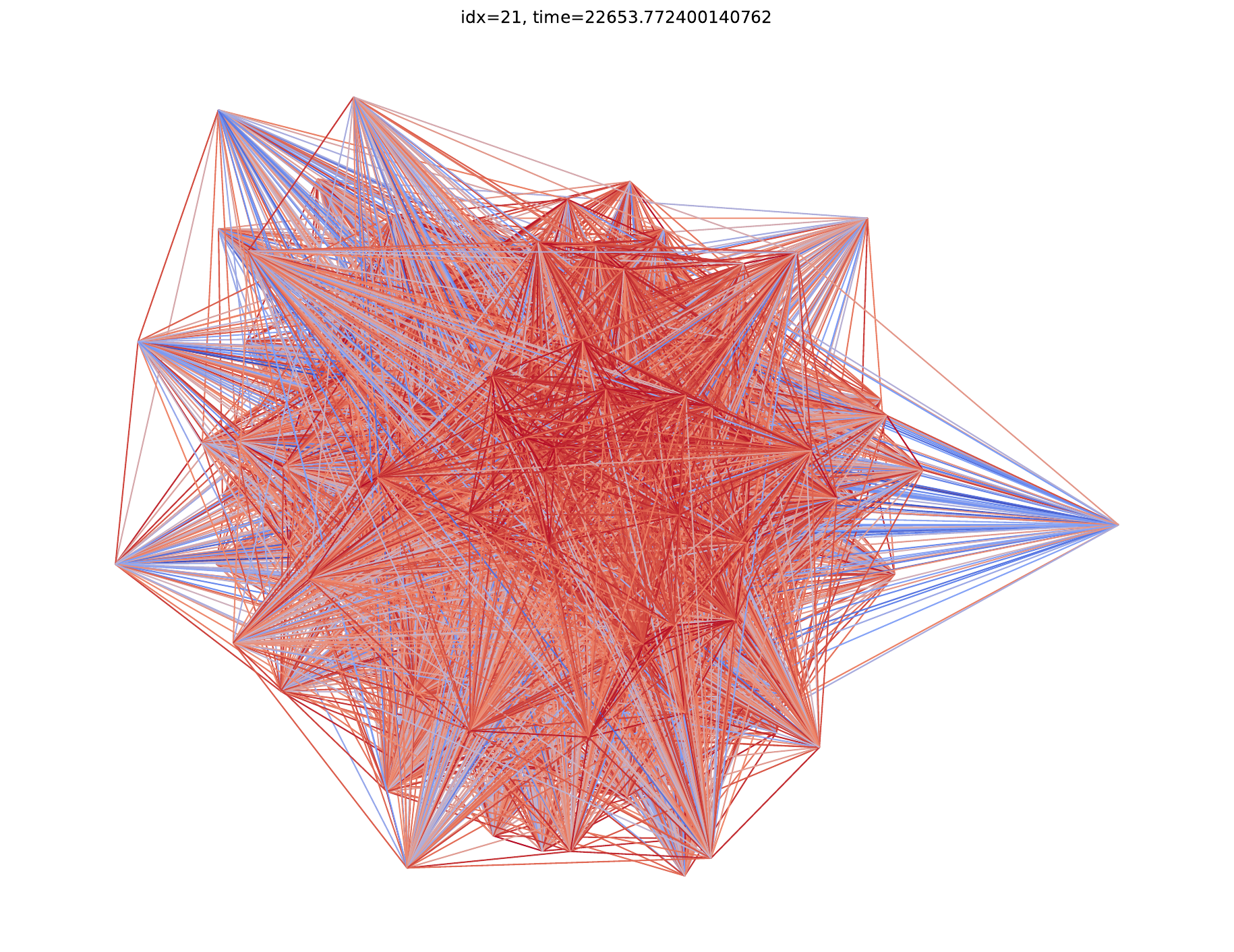} &
\imgcell{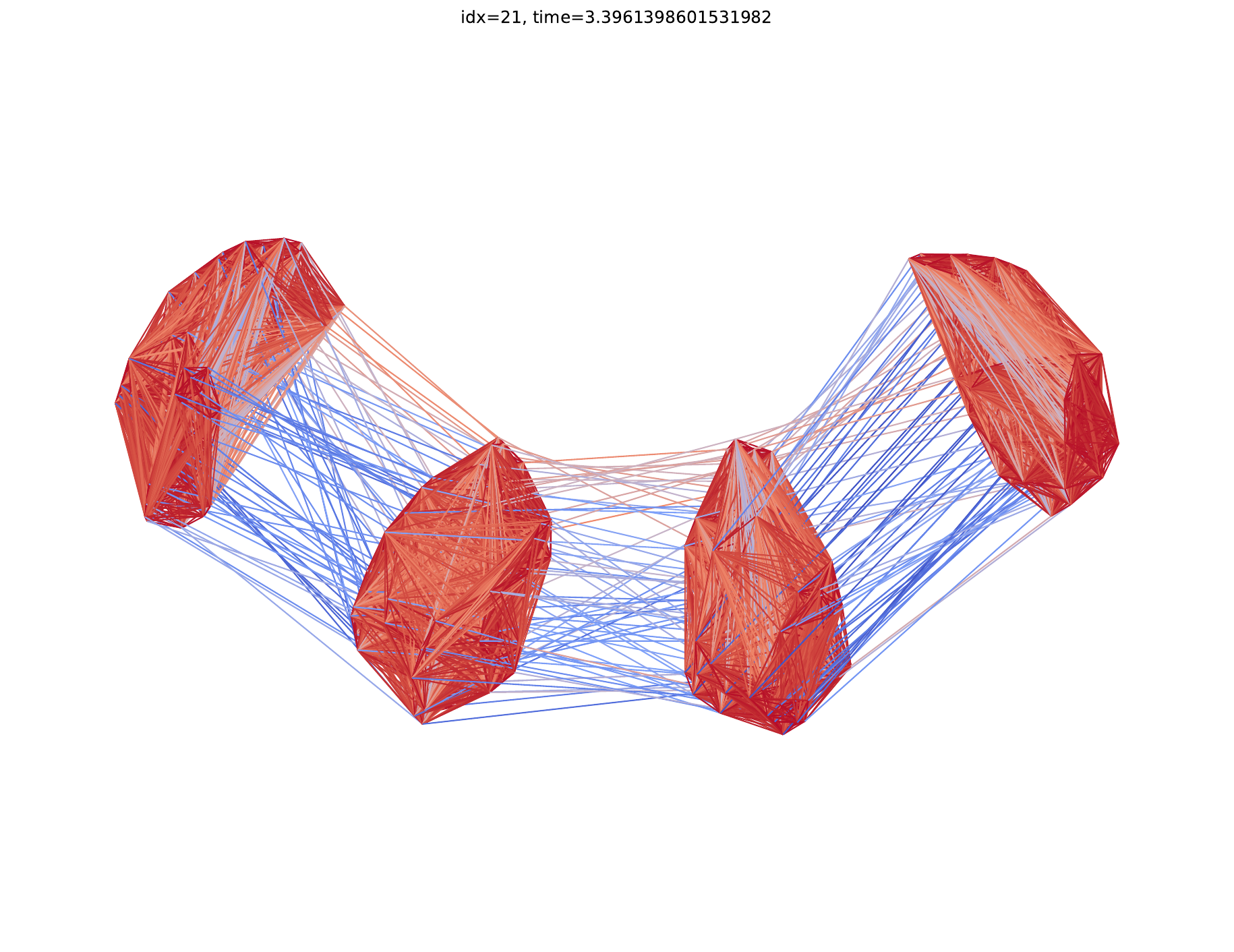} &
\imgcell{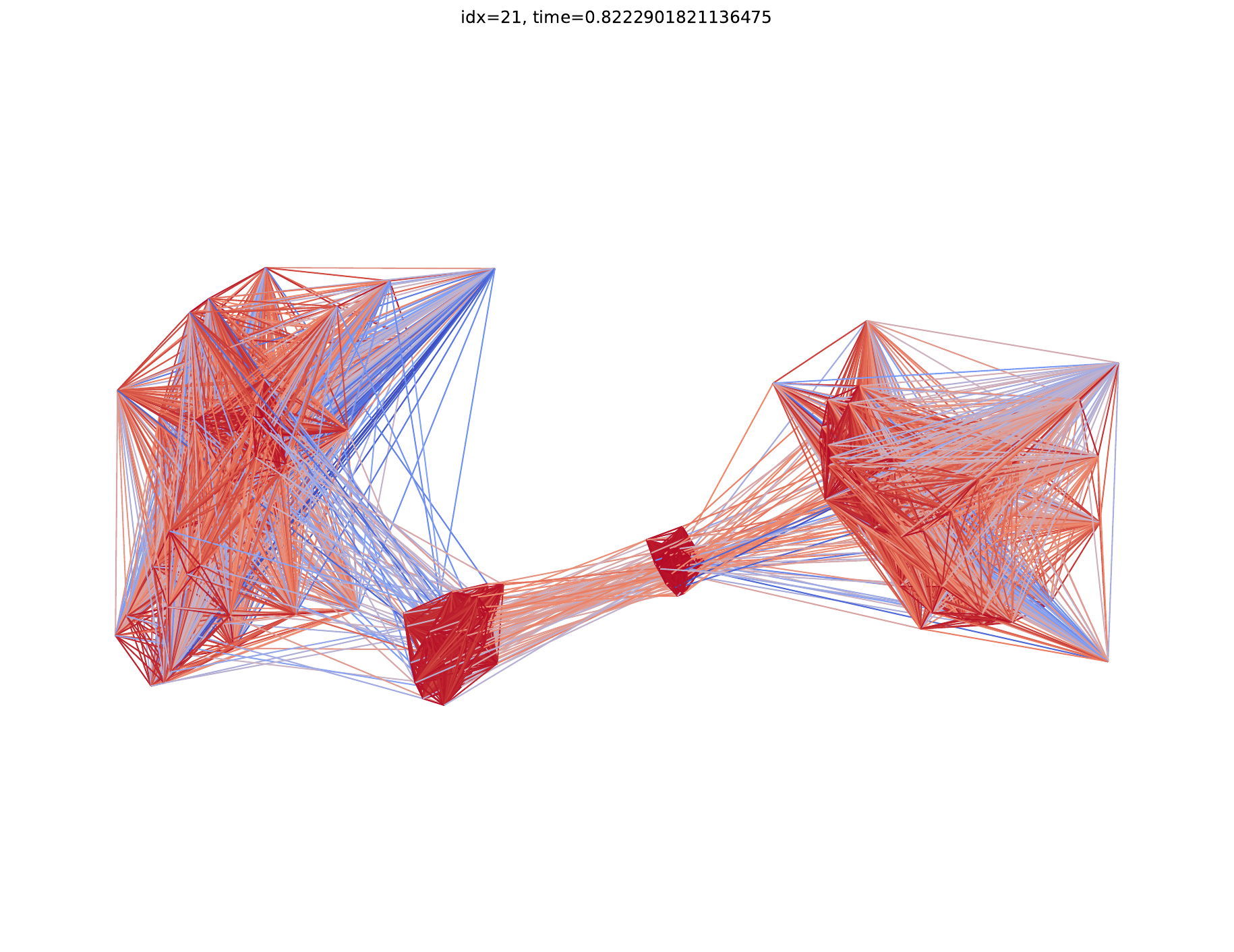} &
\imgcell{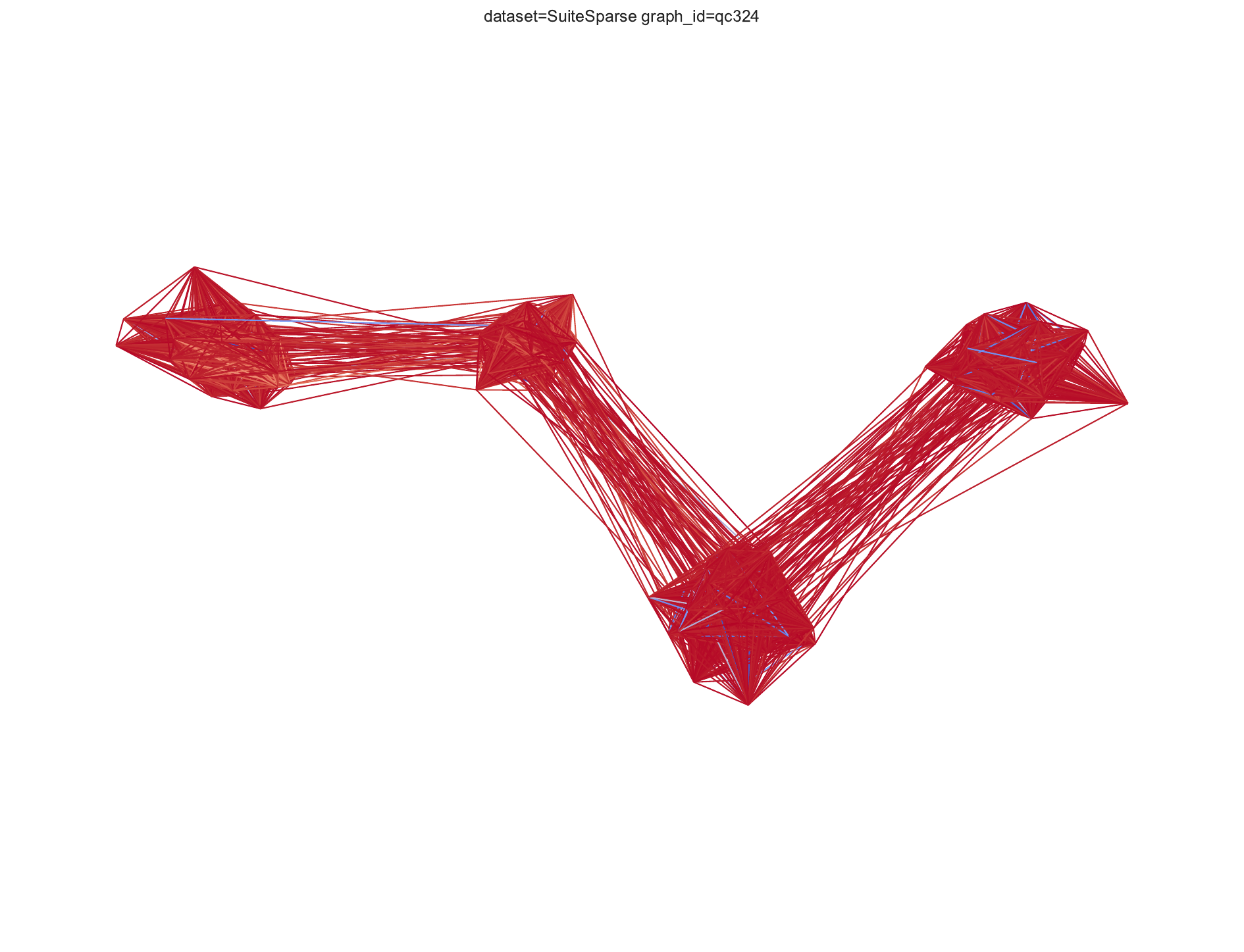} &
\imgcell{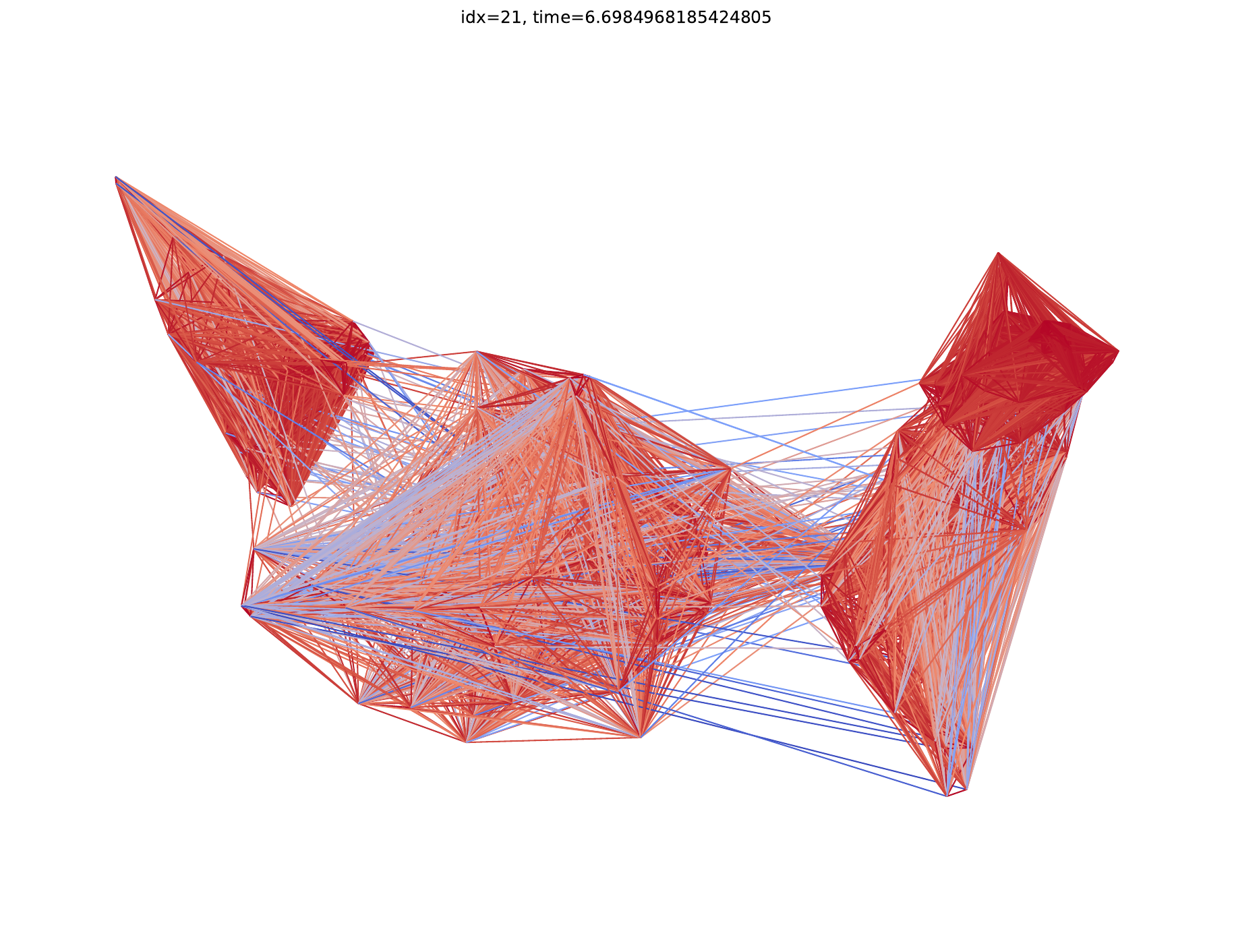} &
\imgcell{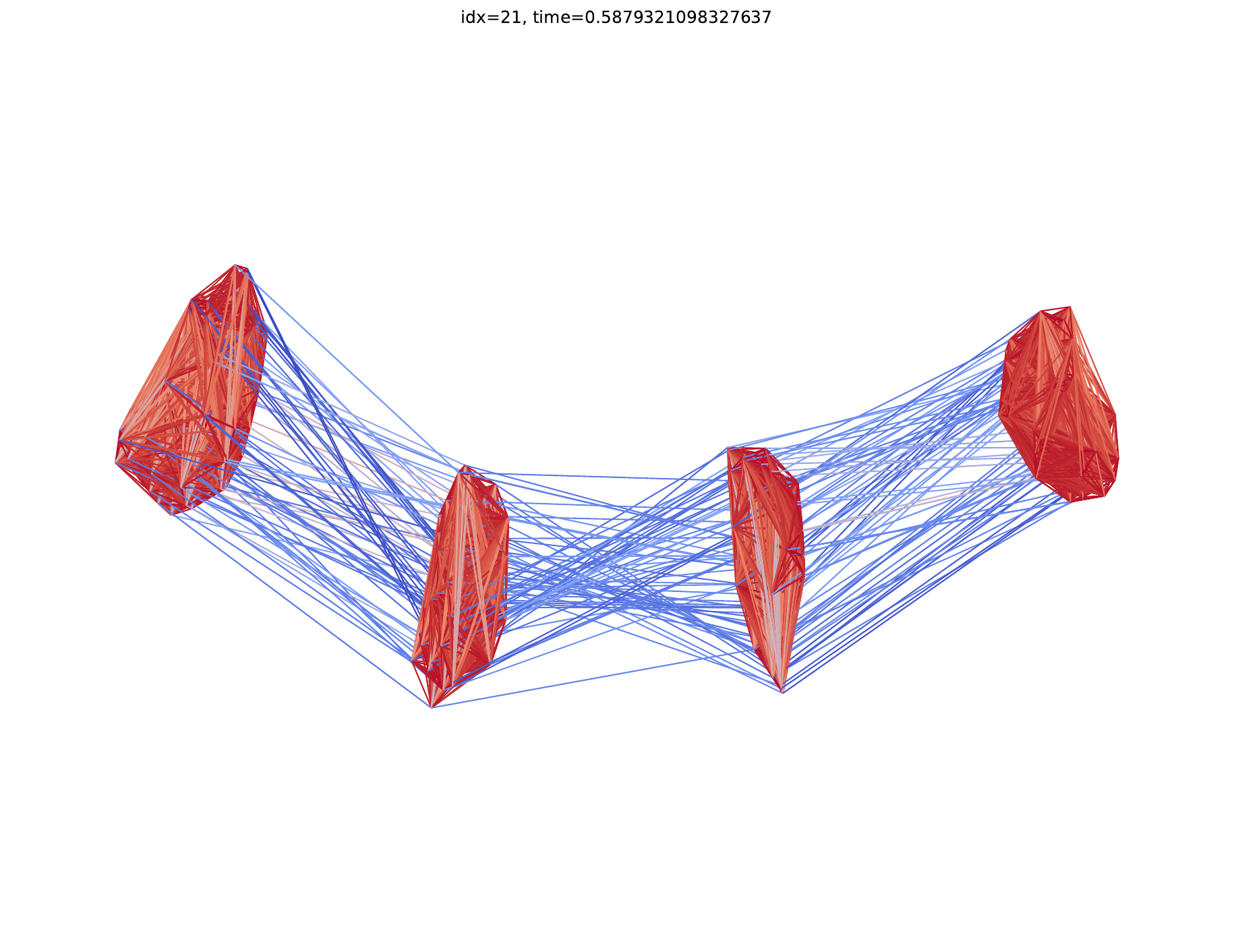} &
\imgcell{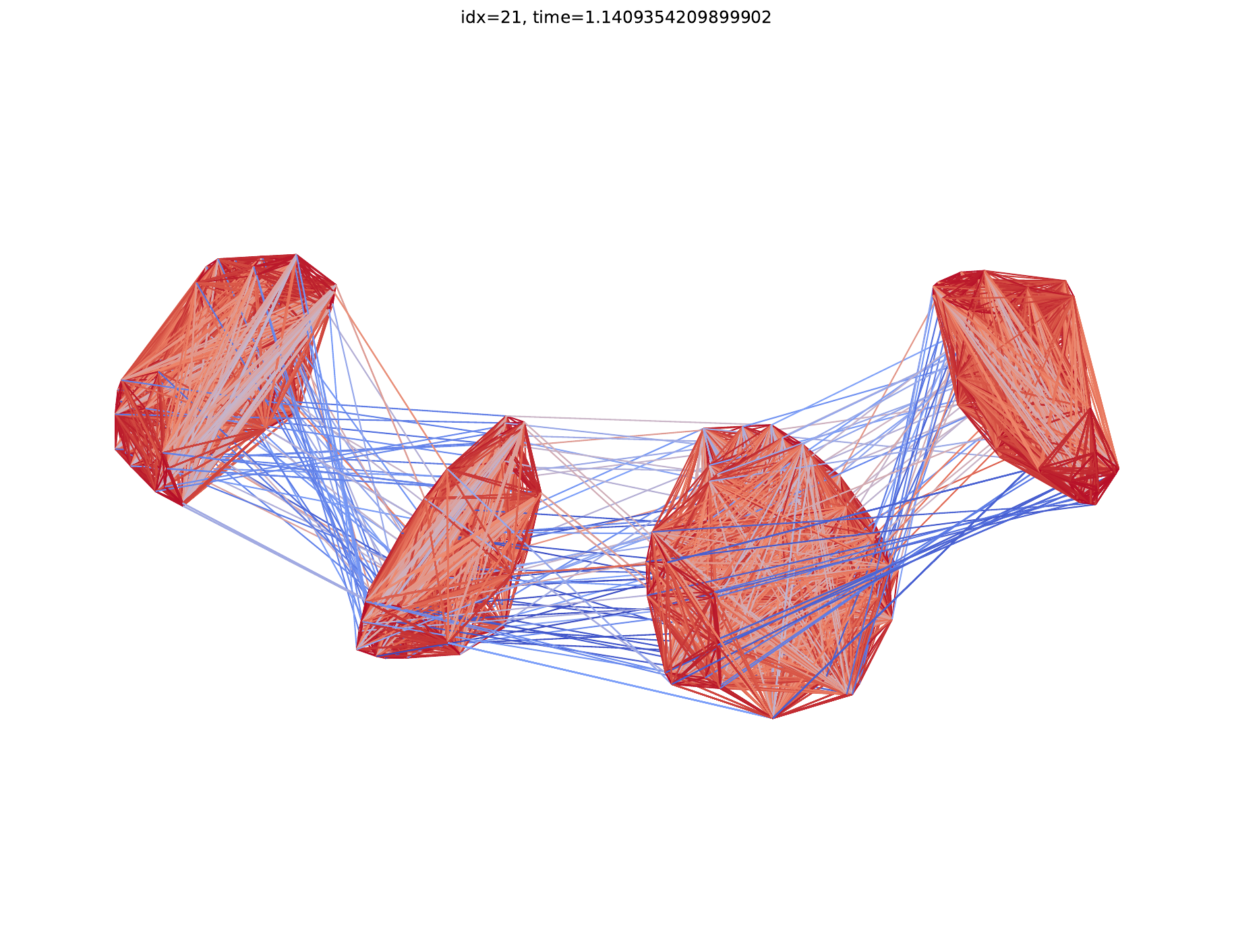} &
\imgcell{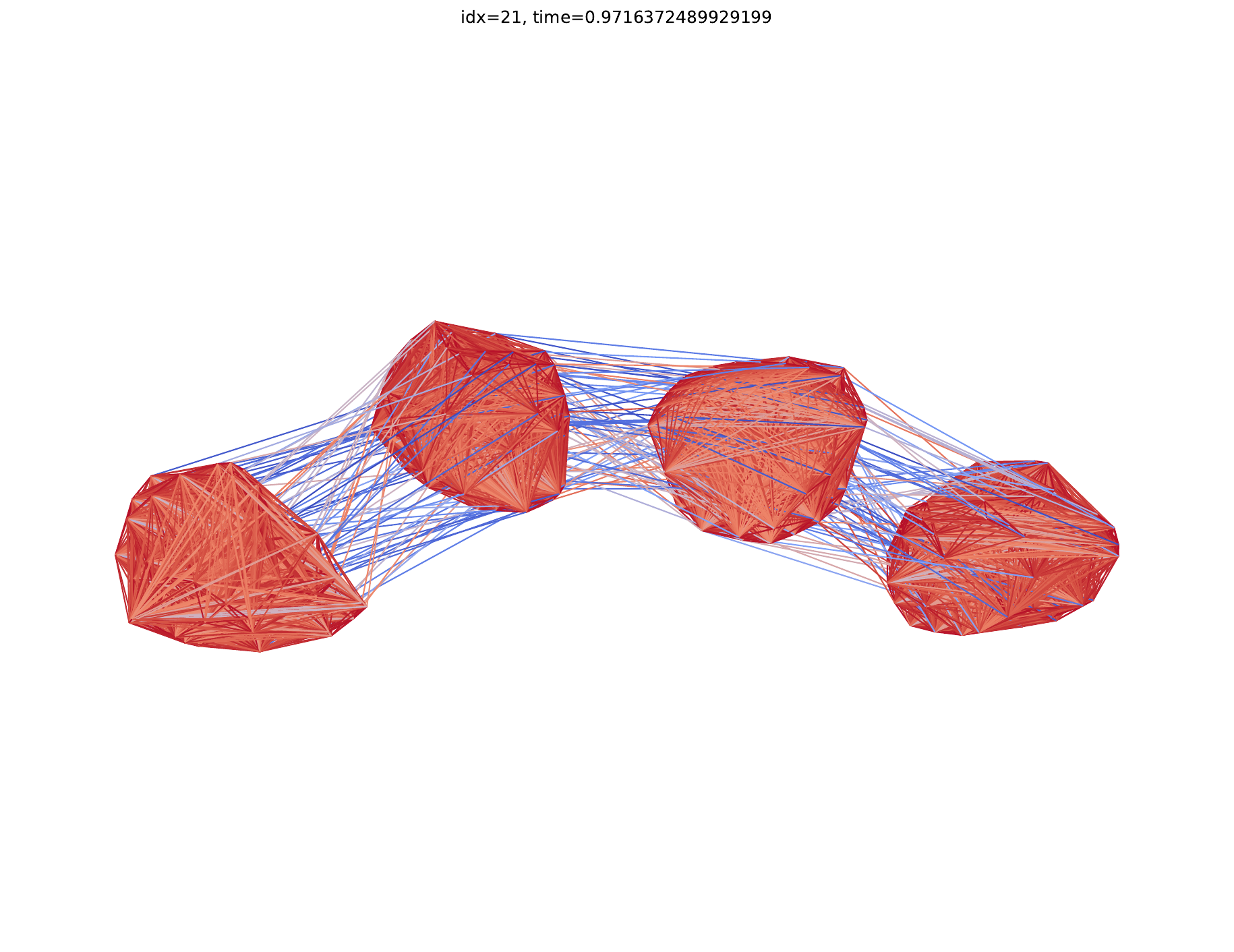} \\

&
t = 0.07s &
t = 44.93s &
t = 10.03s &
t = 0.53s &
t = 7200.00s &
t = 0.41s &
t = 0.46s &
t = 0.56s &
t = 0.44s &
t = 0.58s &
t = 0.49s &
t = 0.48s \\

\makecell{\bfseries Ecoli\_10NN\\N = 336\\M = 2280} &
\imgcell{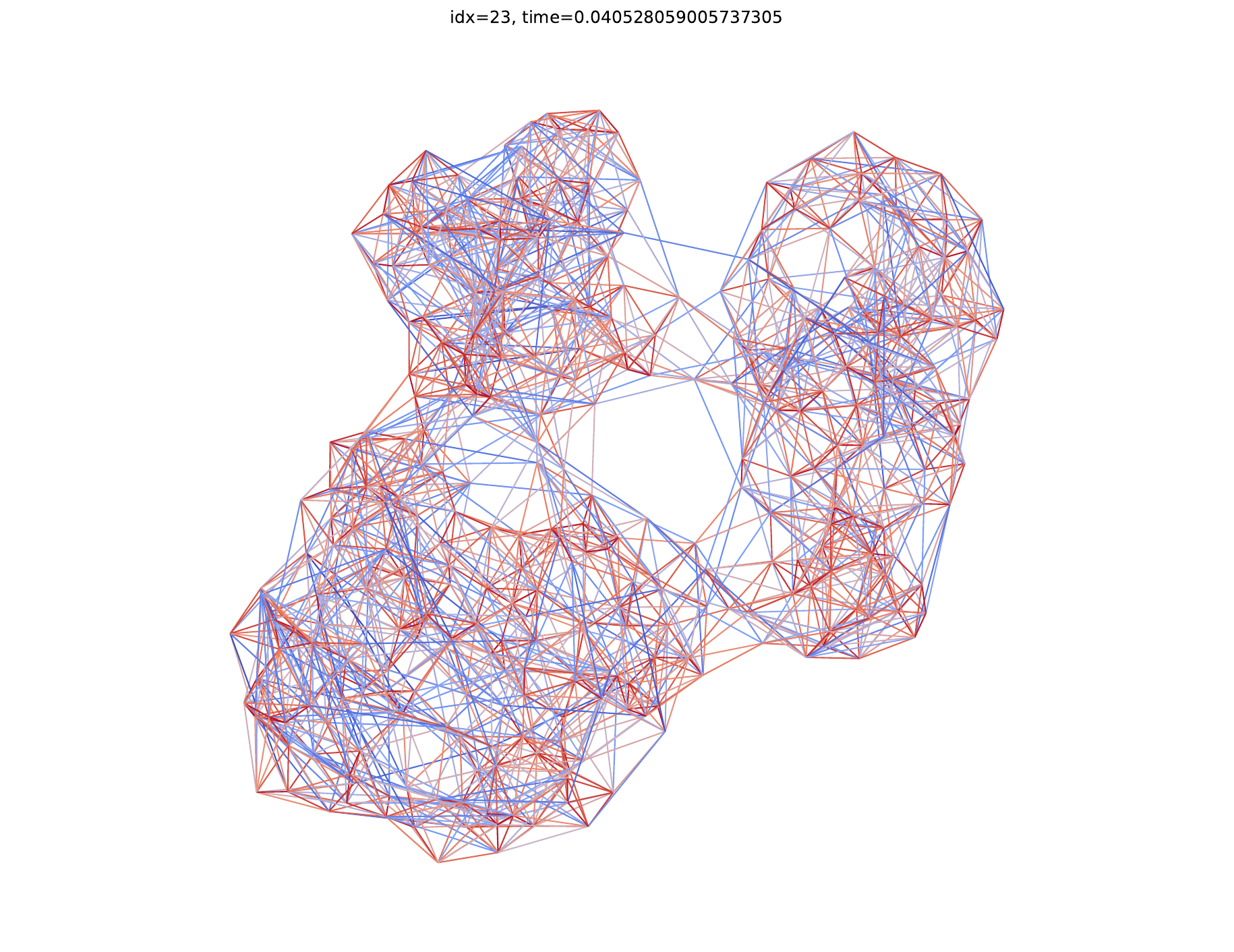} &
\imgcell{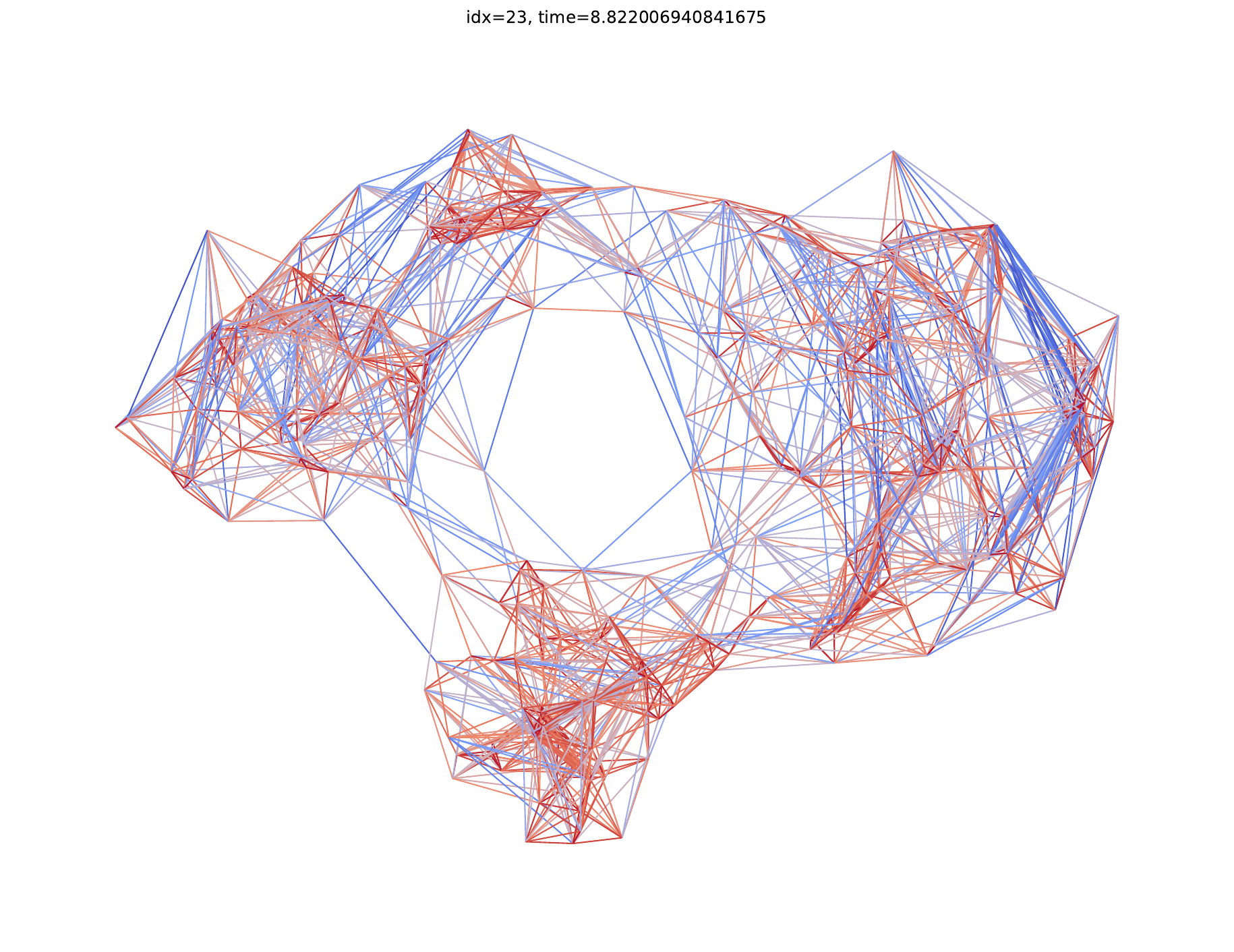} &
\imgcell{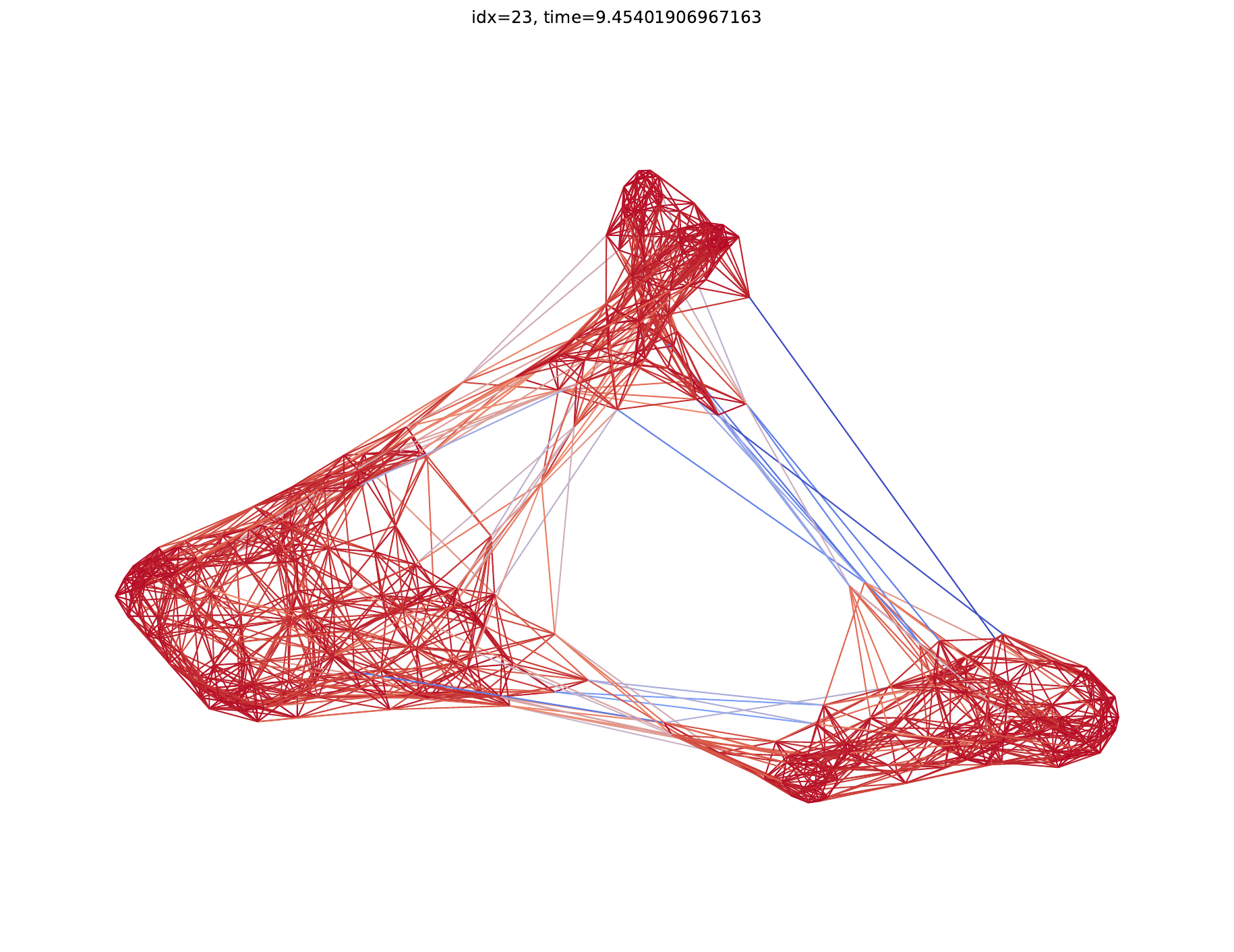} &
\imgcell{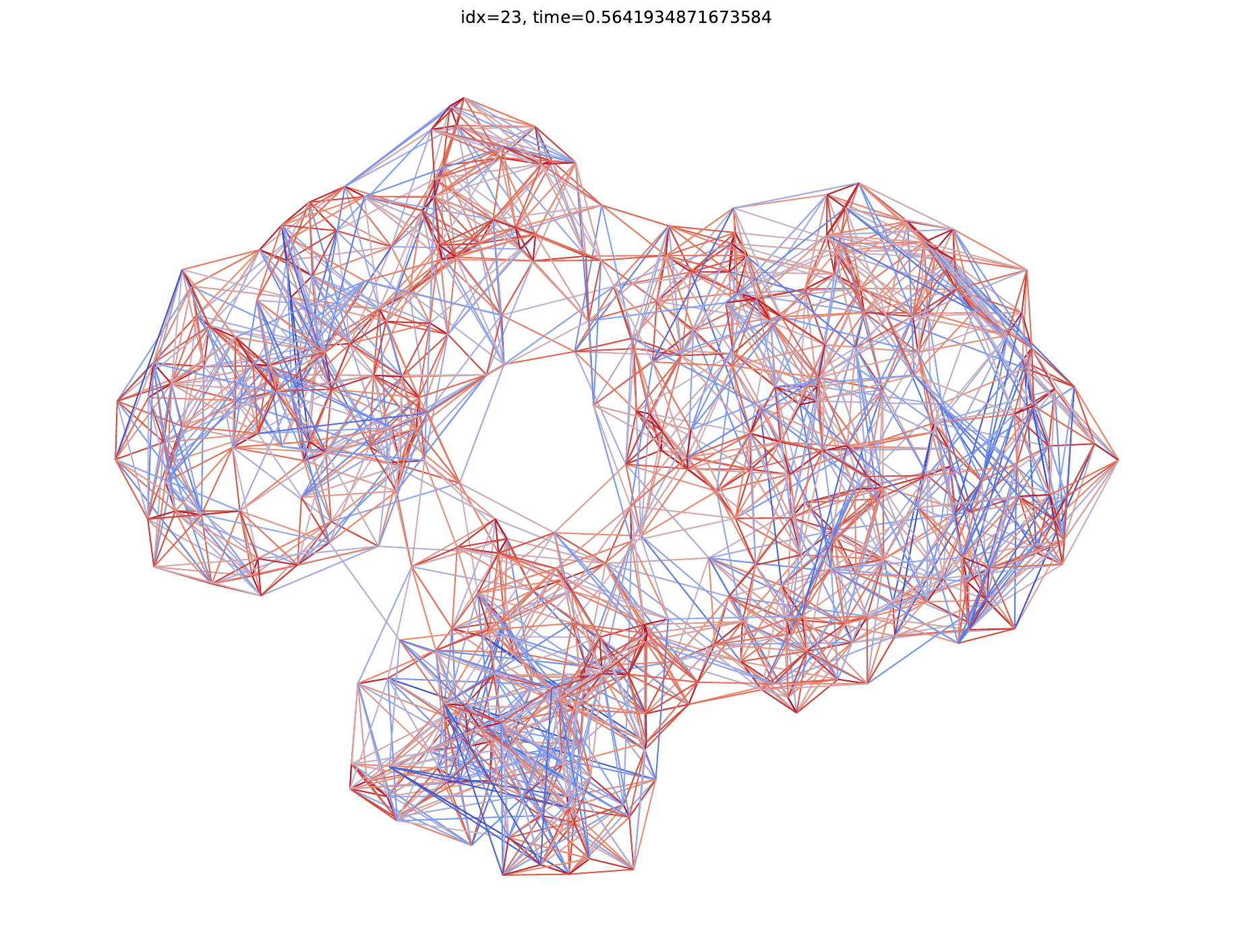} &
\imgcell{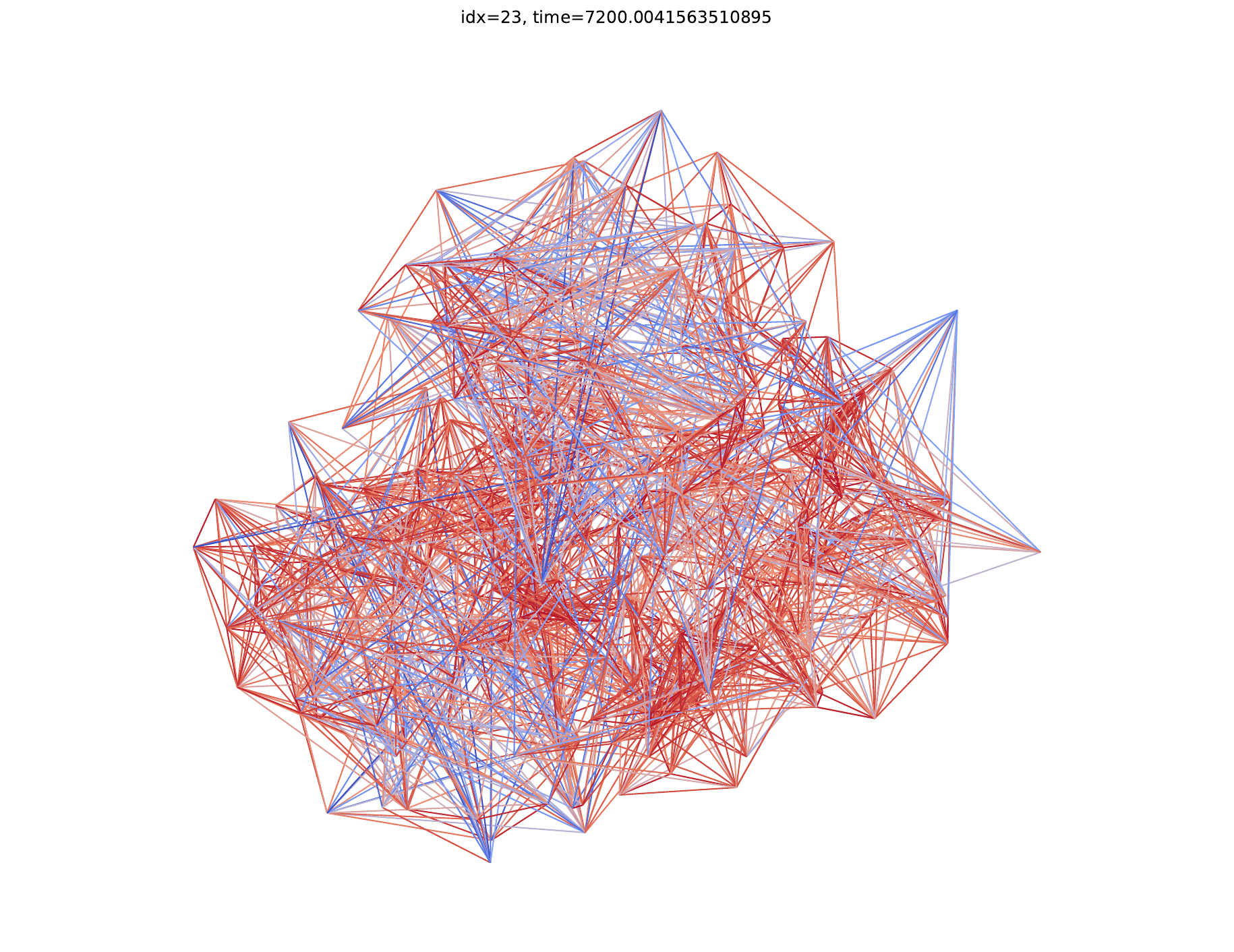} &
\imgcell{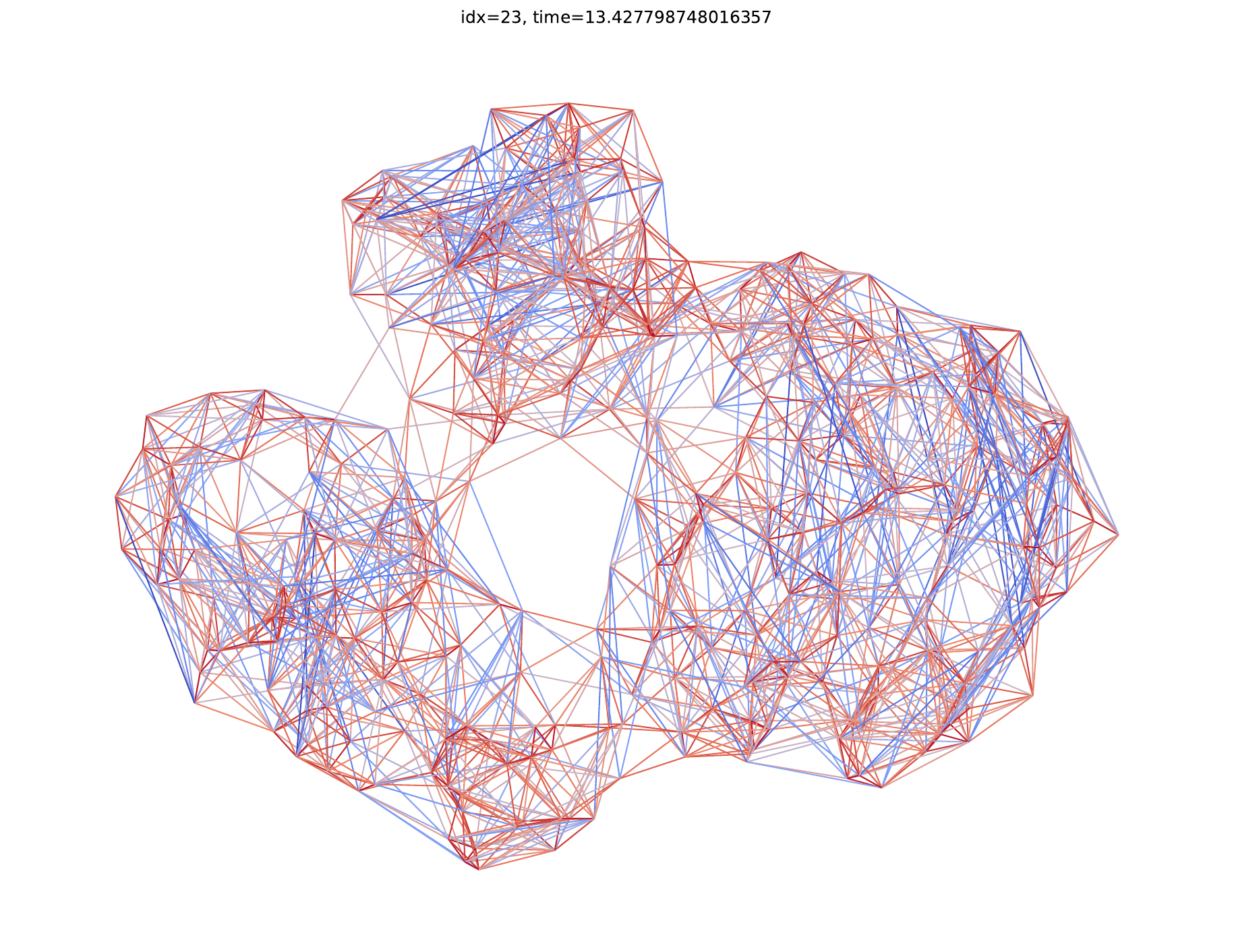} &
\imgcell{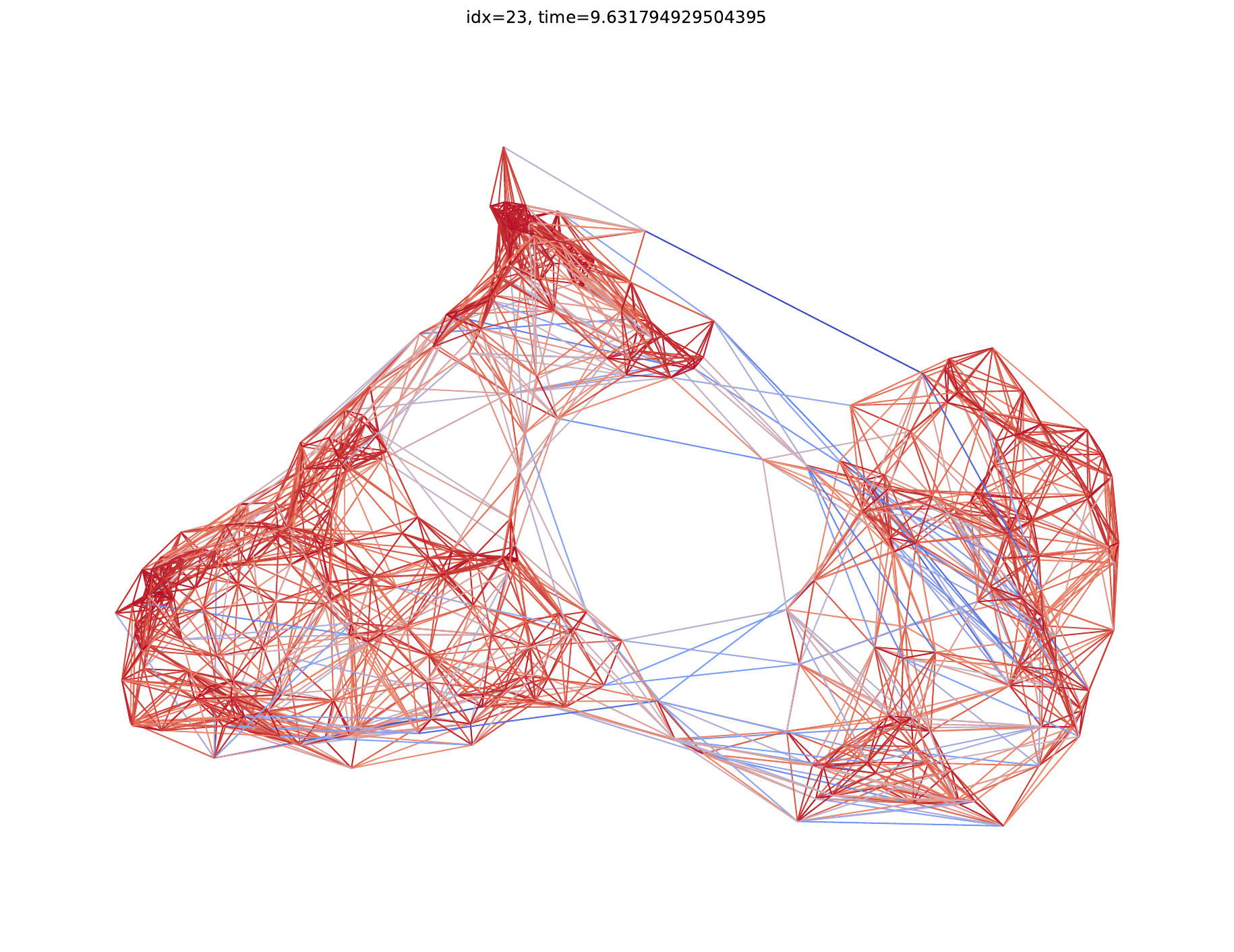} &
\imgcell{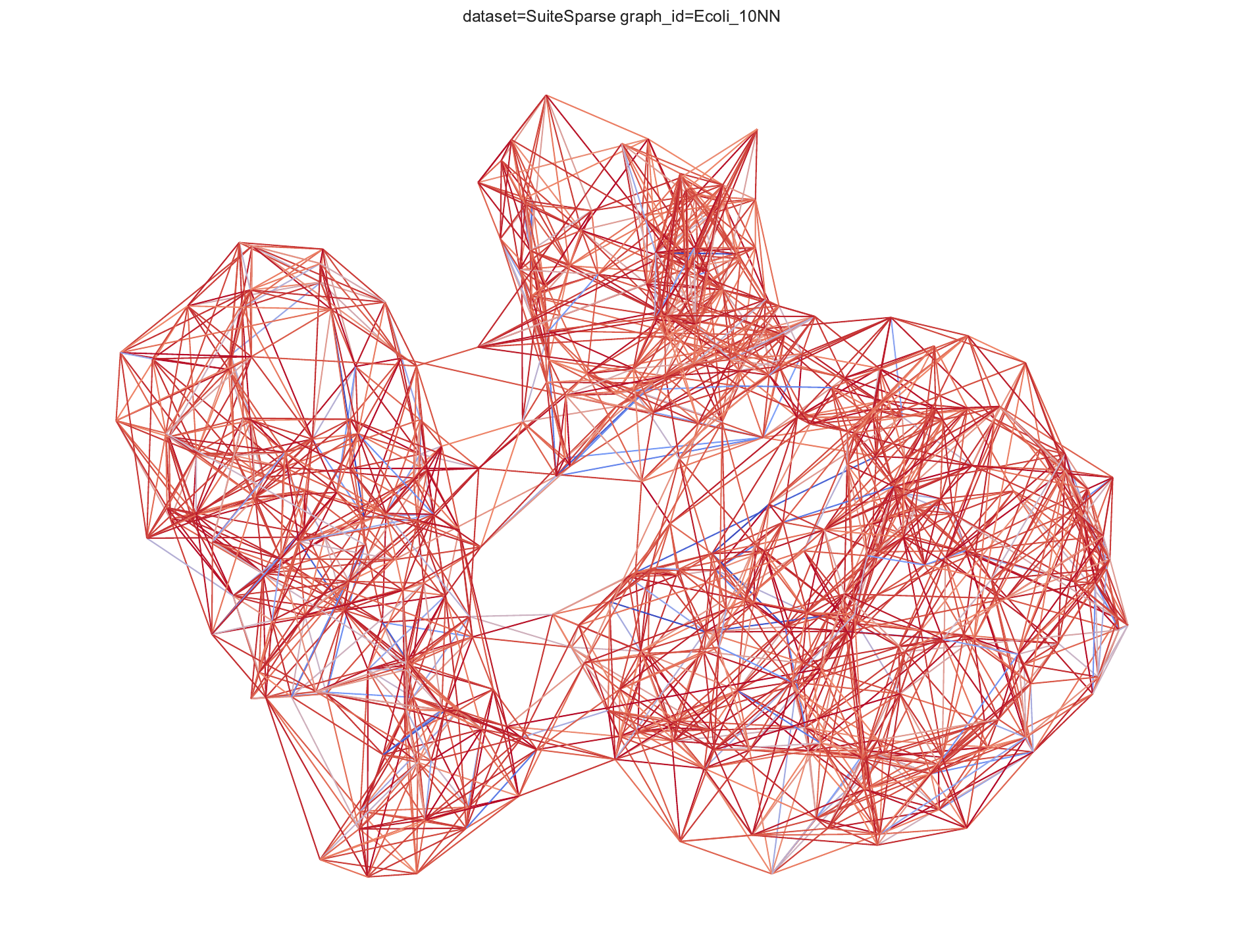} &
\imgcell{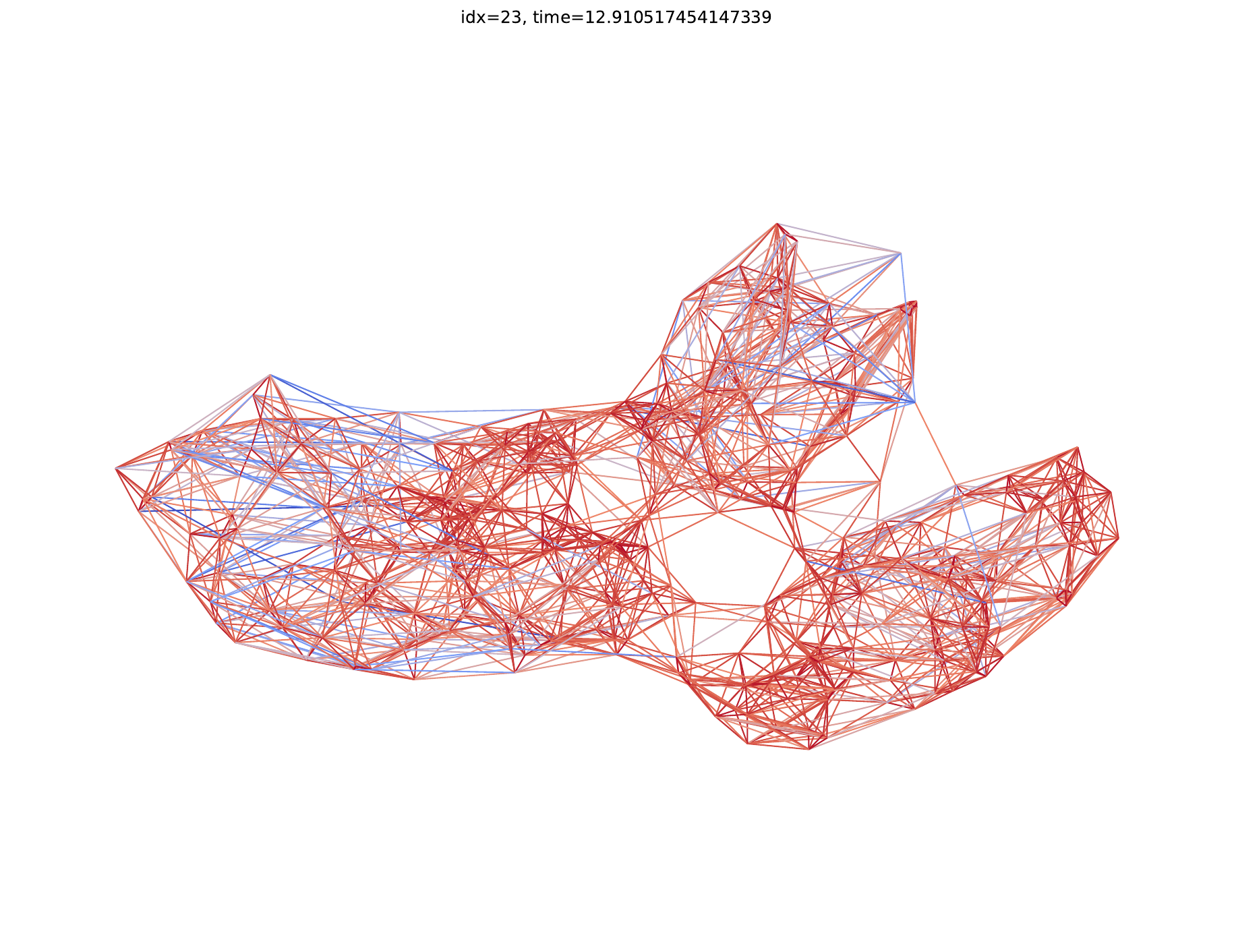} &
\imgcell{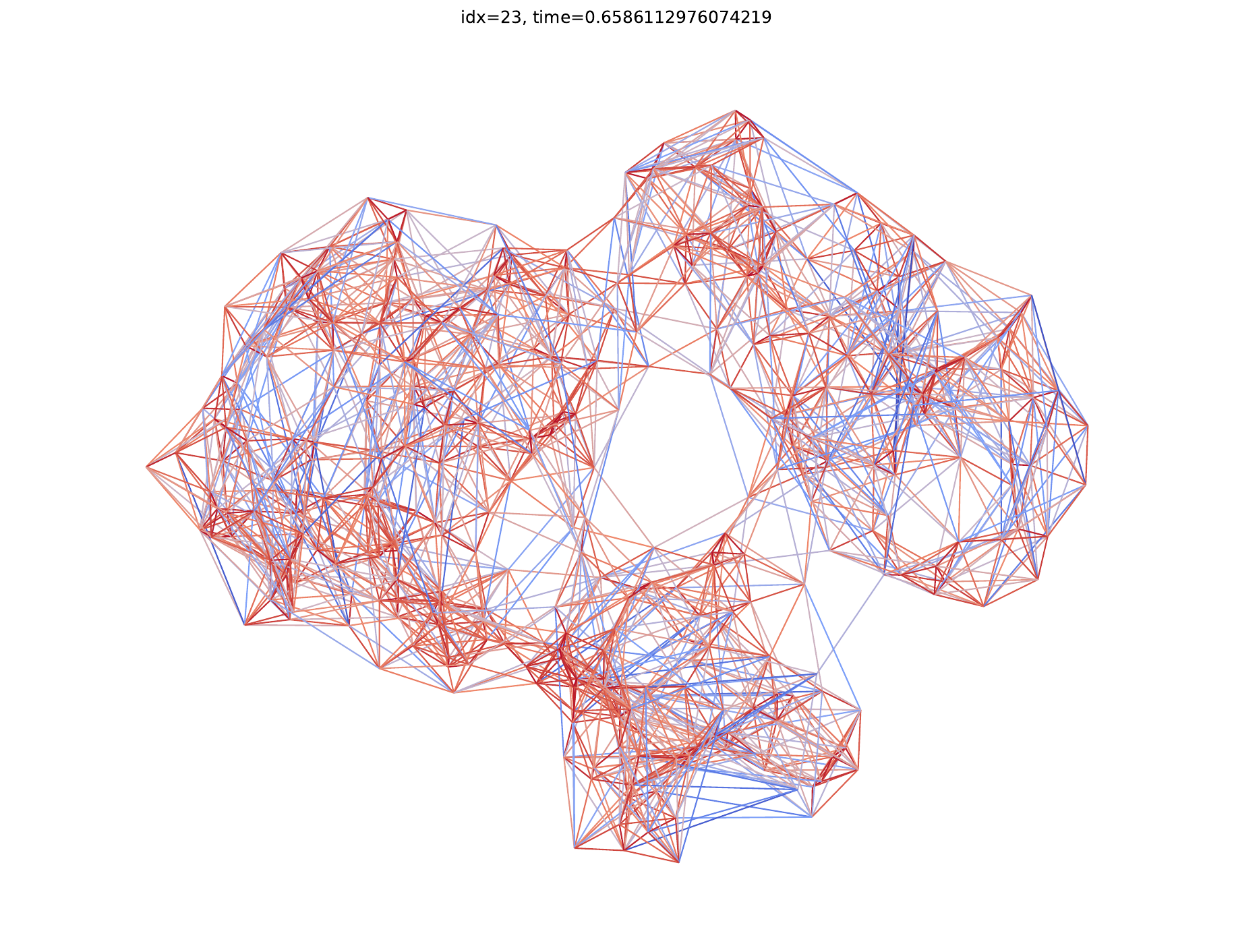} &
\imgcell{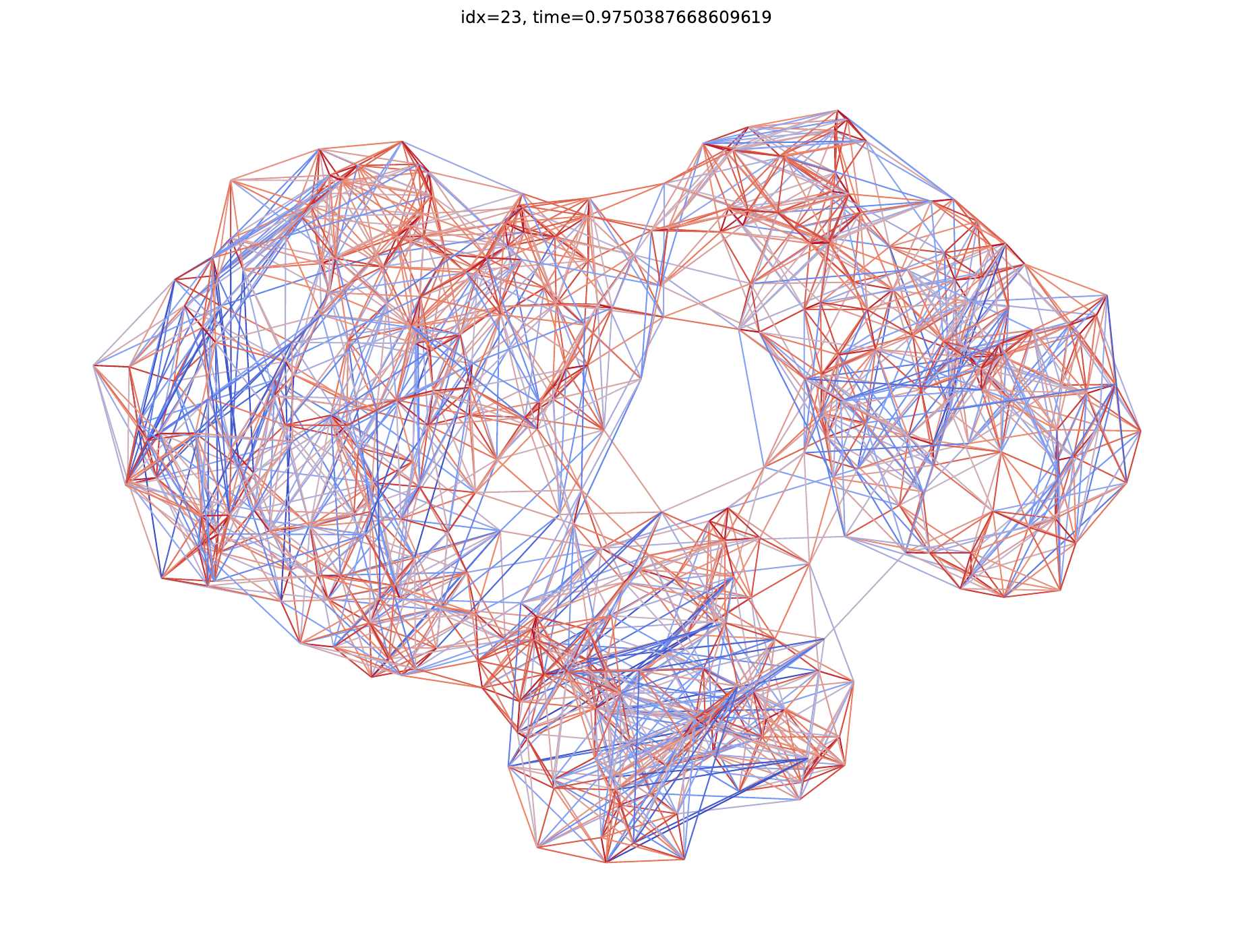} &
\imgcell{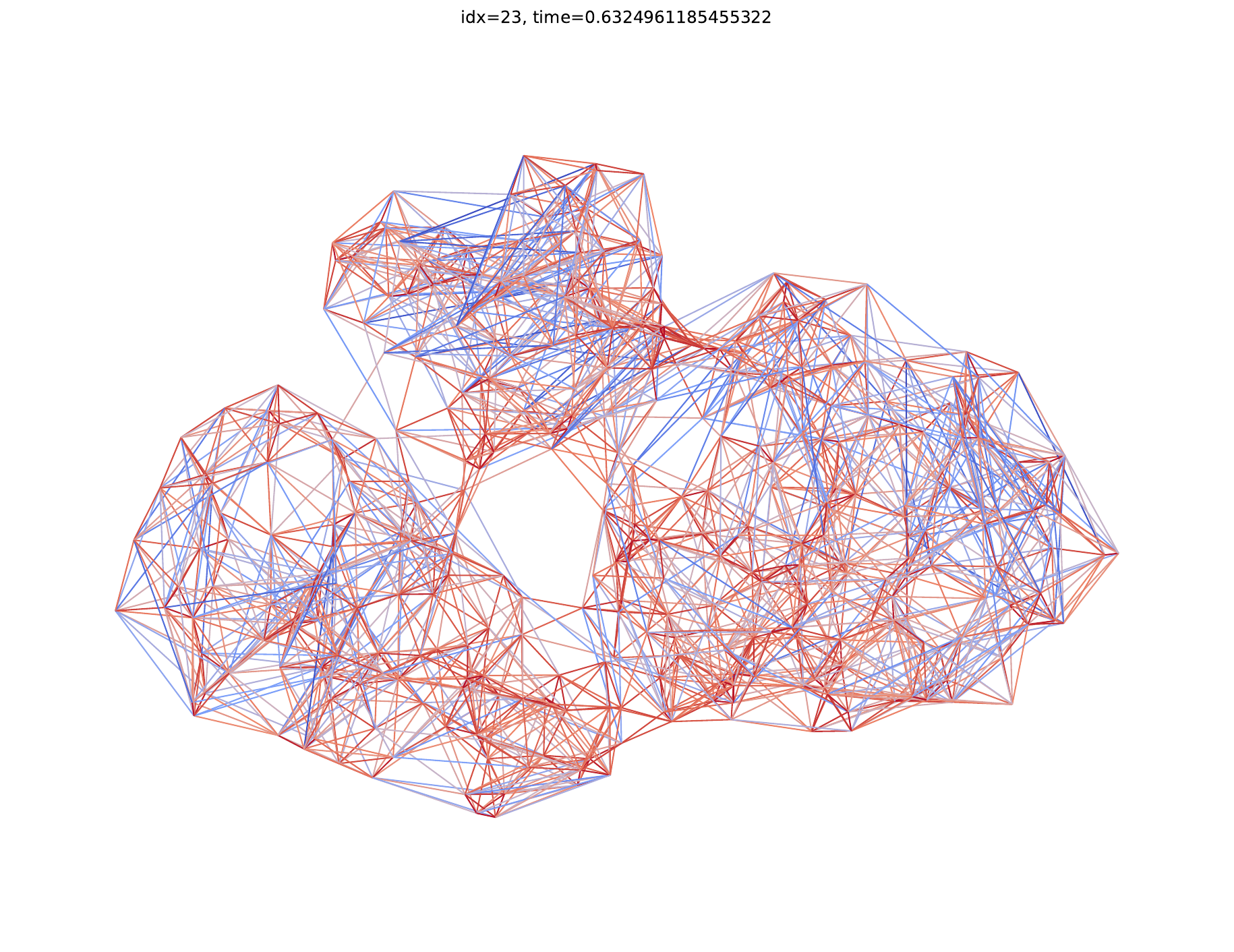} \\

&
t = 0.04s &
t = 8.82s &
t = 9.45s &
t = 0.56s &
t = 7200.00s &
t = 0.67s &
t = 0.42s &
t = 0.47s &
t = 0.37s &
t = 0.48s &
t = 0.51s &
t = 0.61s \\

\makecell{\bfseries cage7\\N = 340\\M = 1372} &
\imgcell{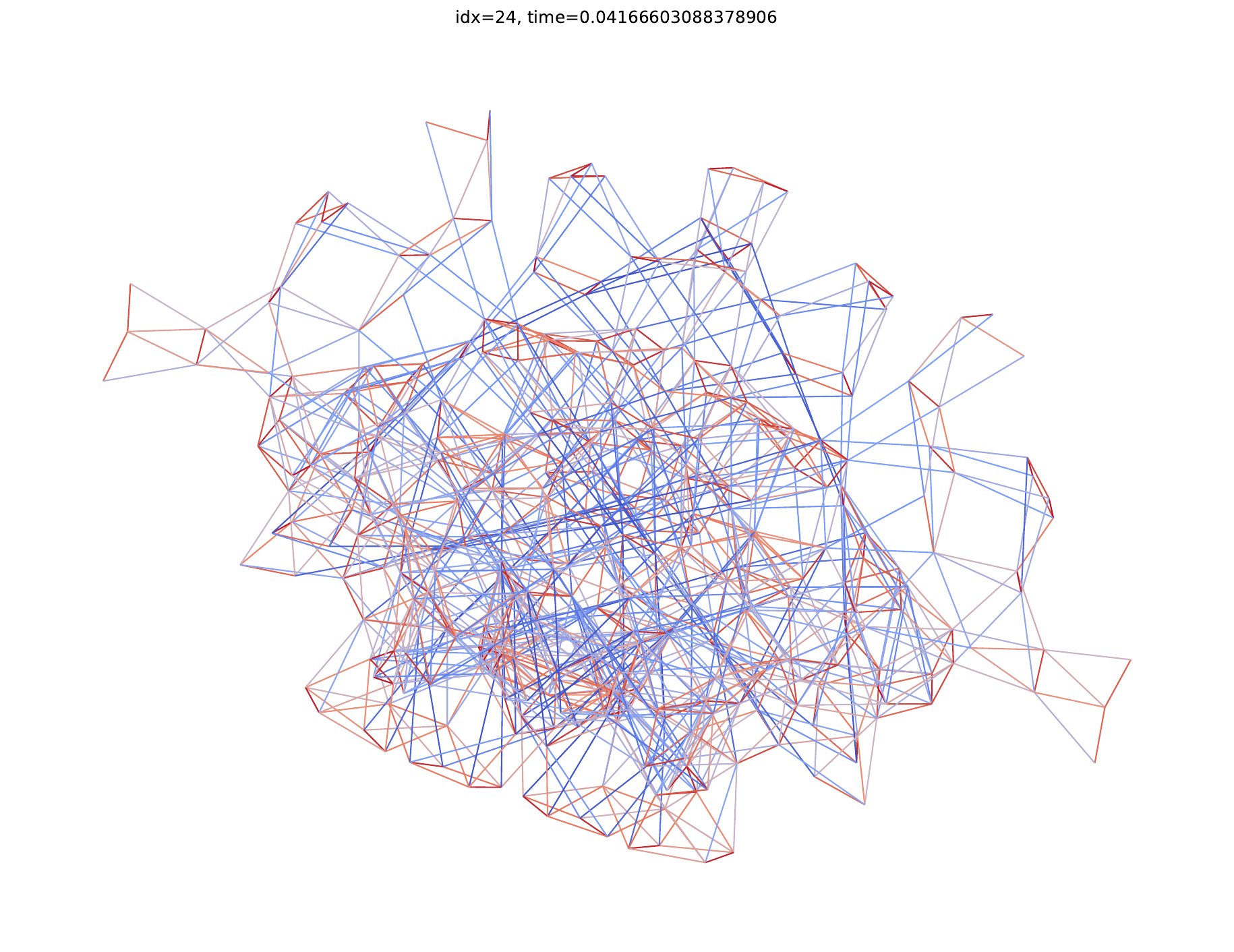} &
\imgcell{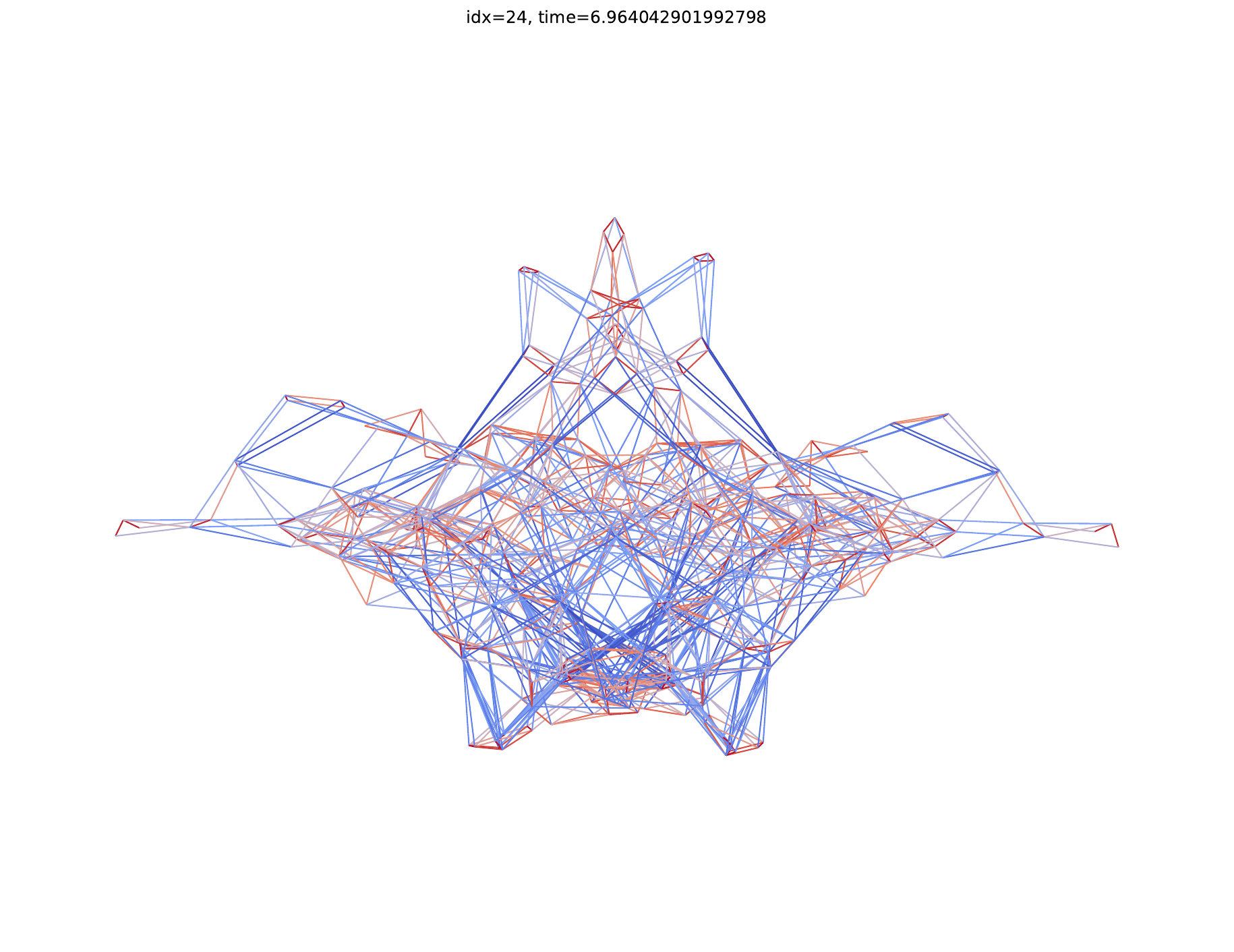} &
\imgcell{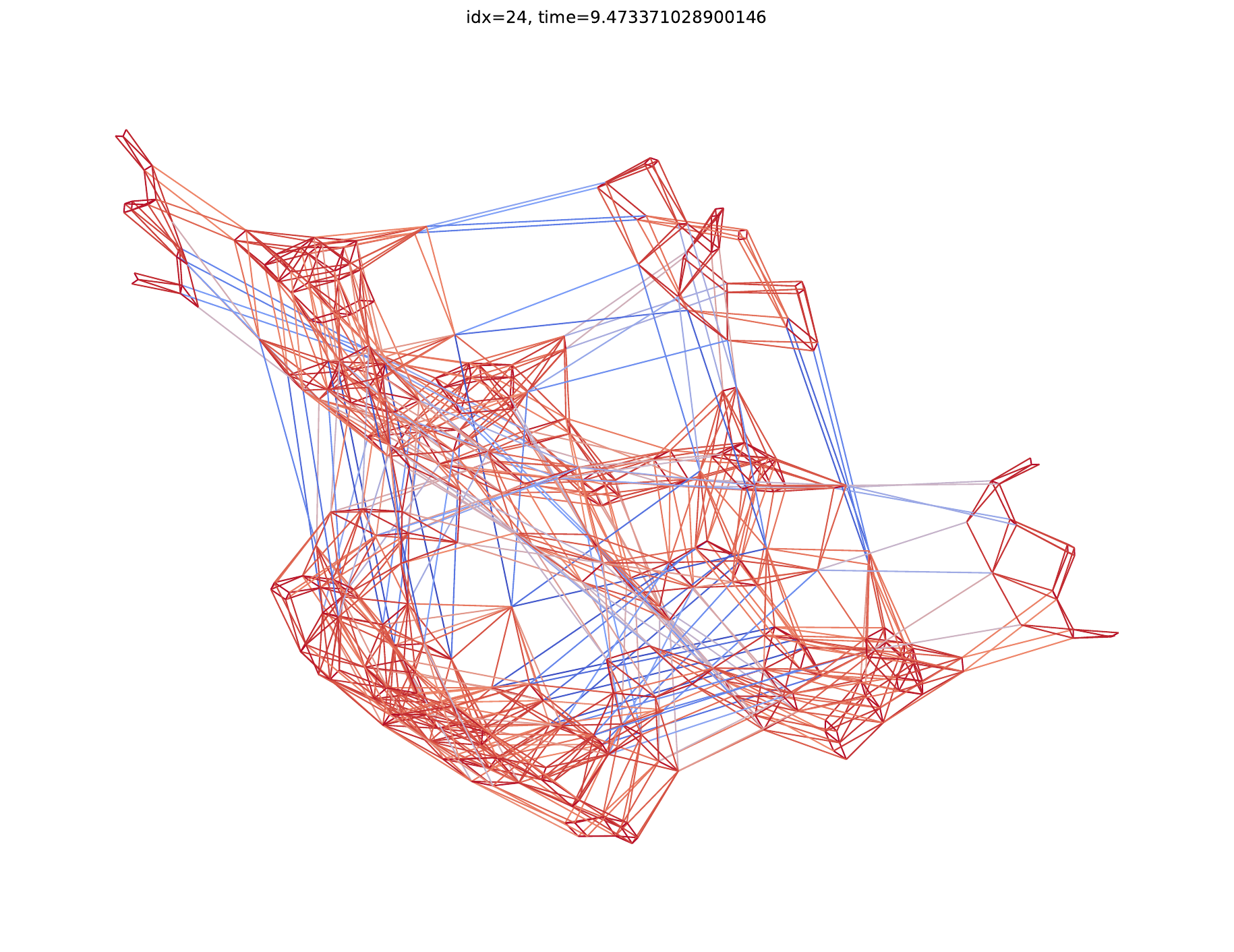} &
\imgcell{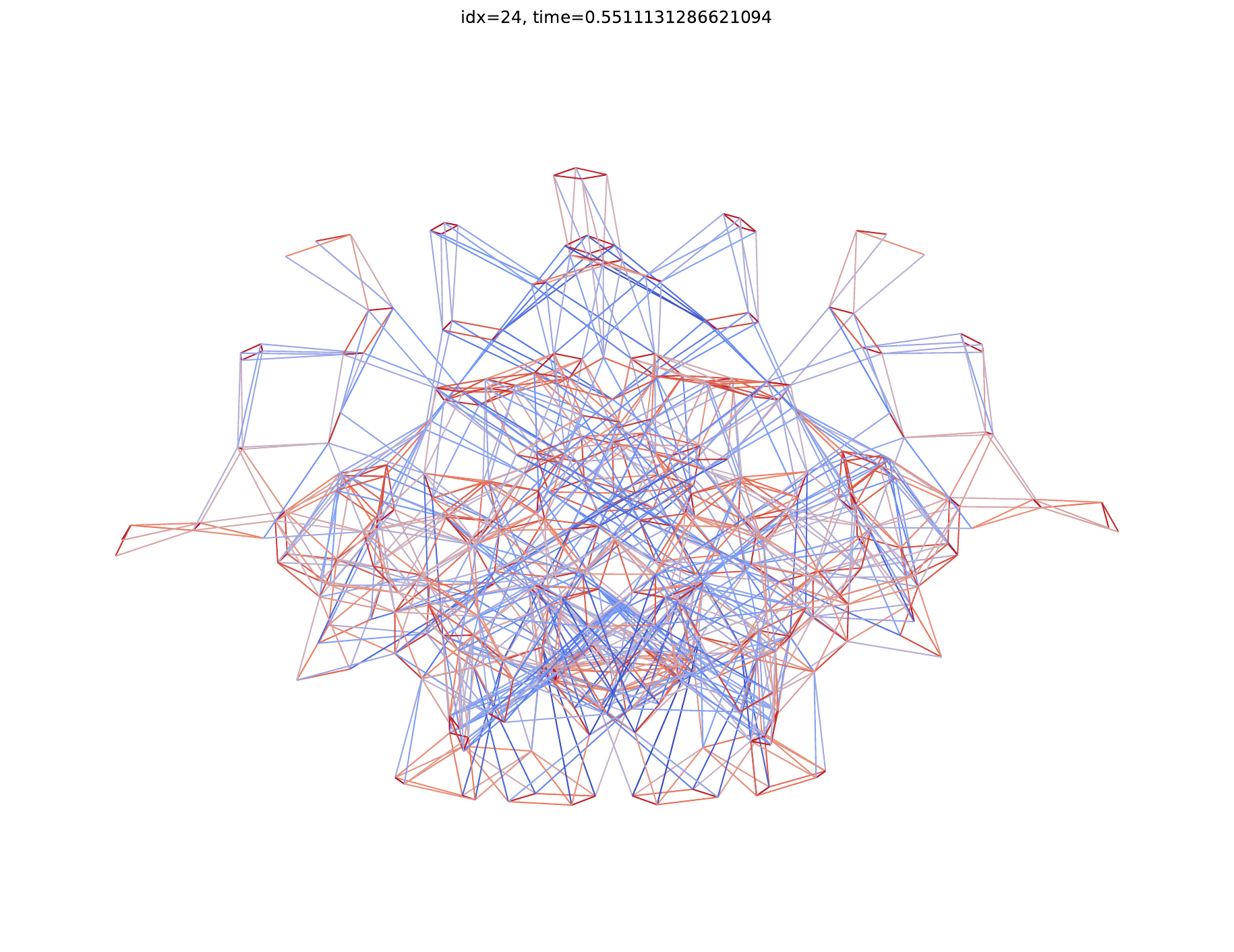} &
\imgcell{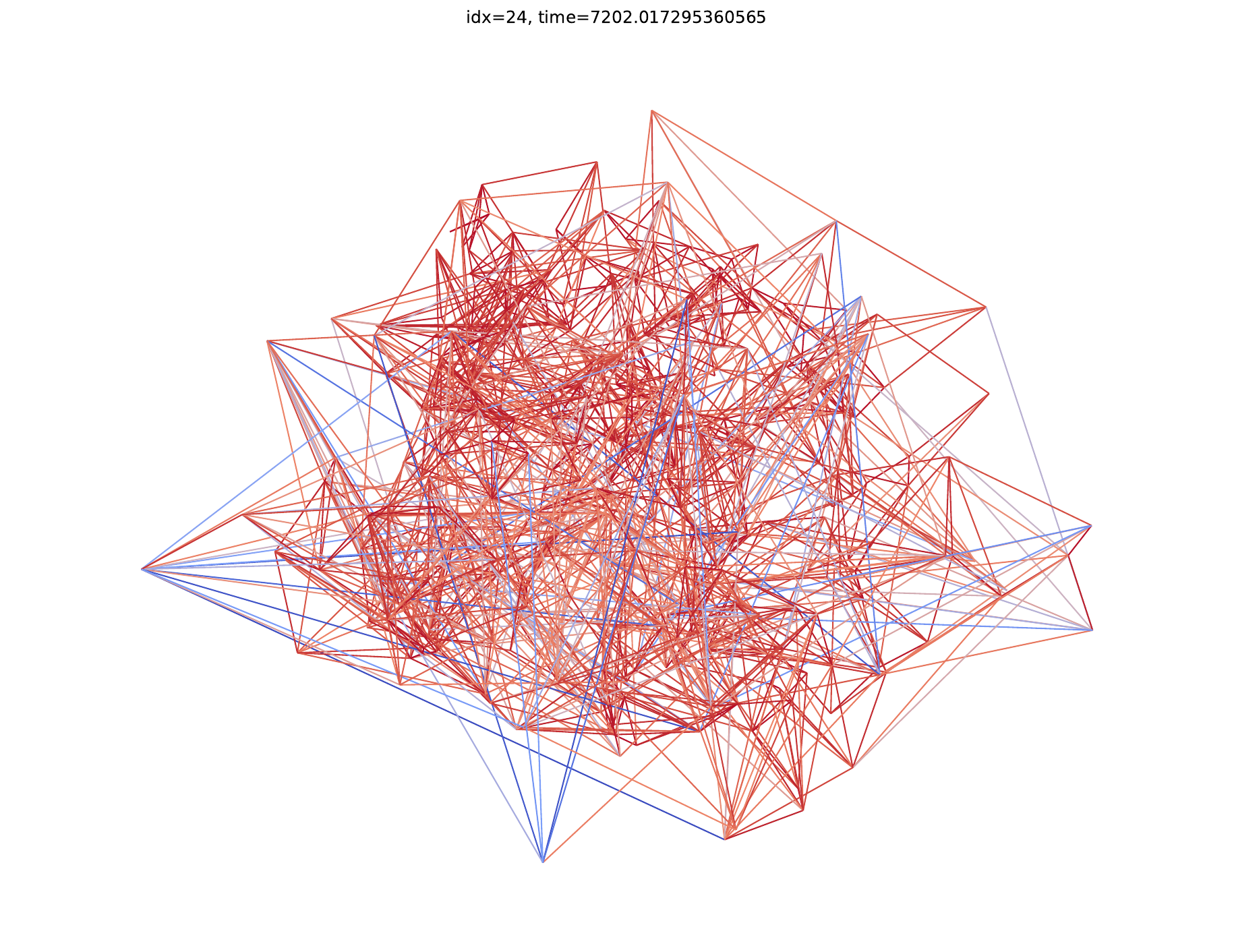} &
\imgcell{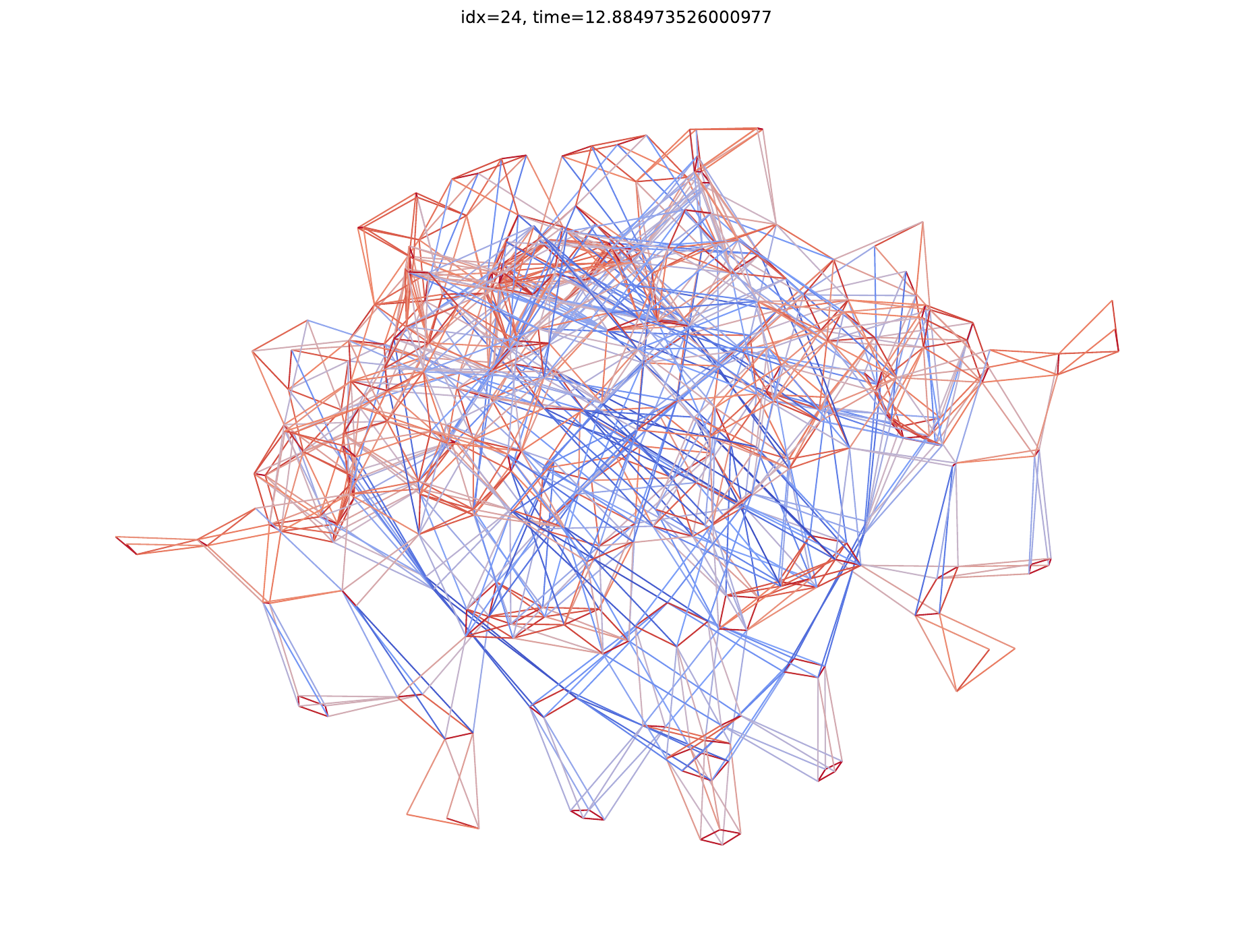} &
\imgcell{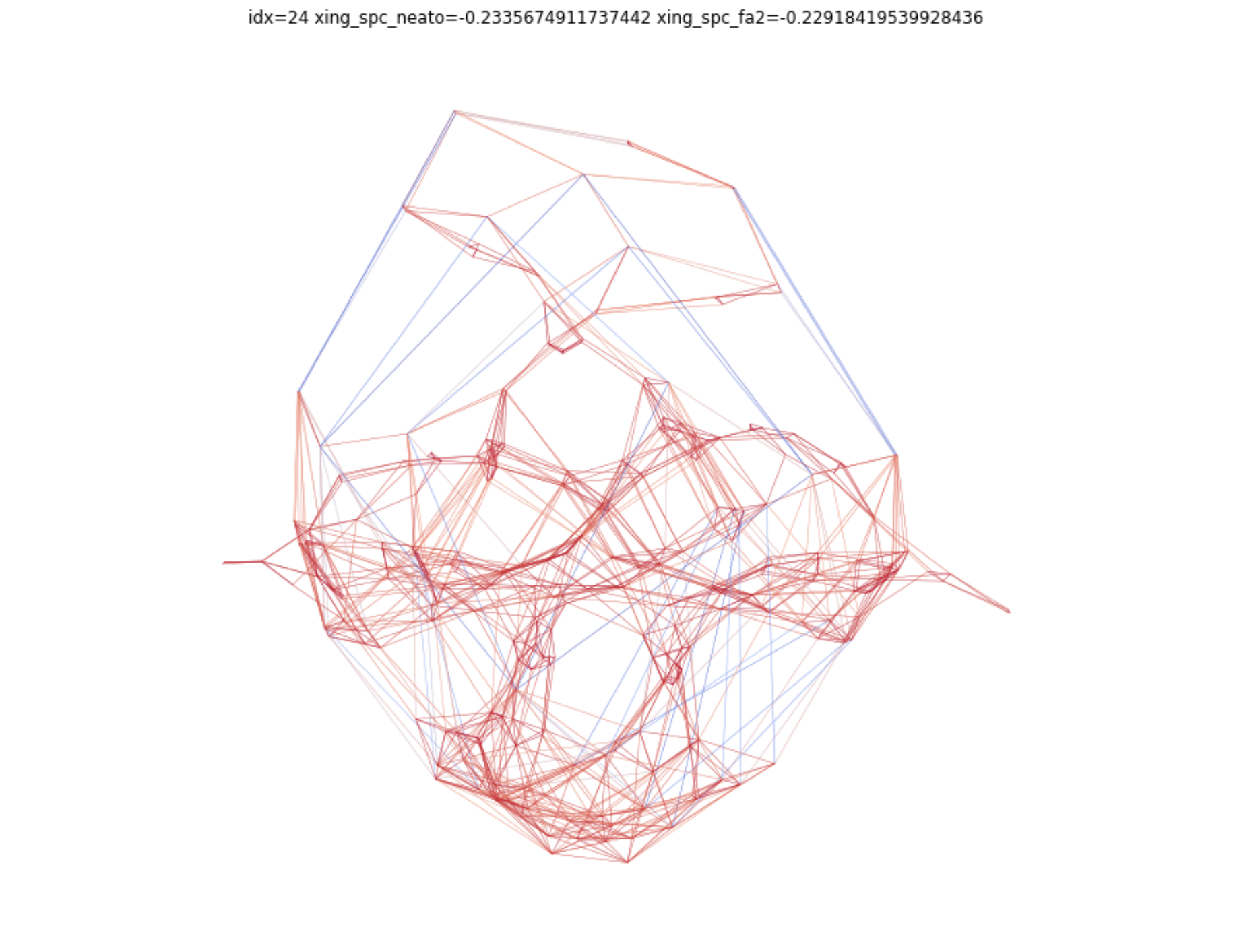} &
\imgcell{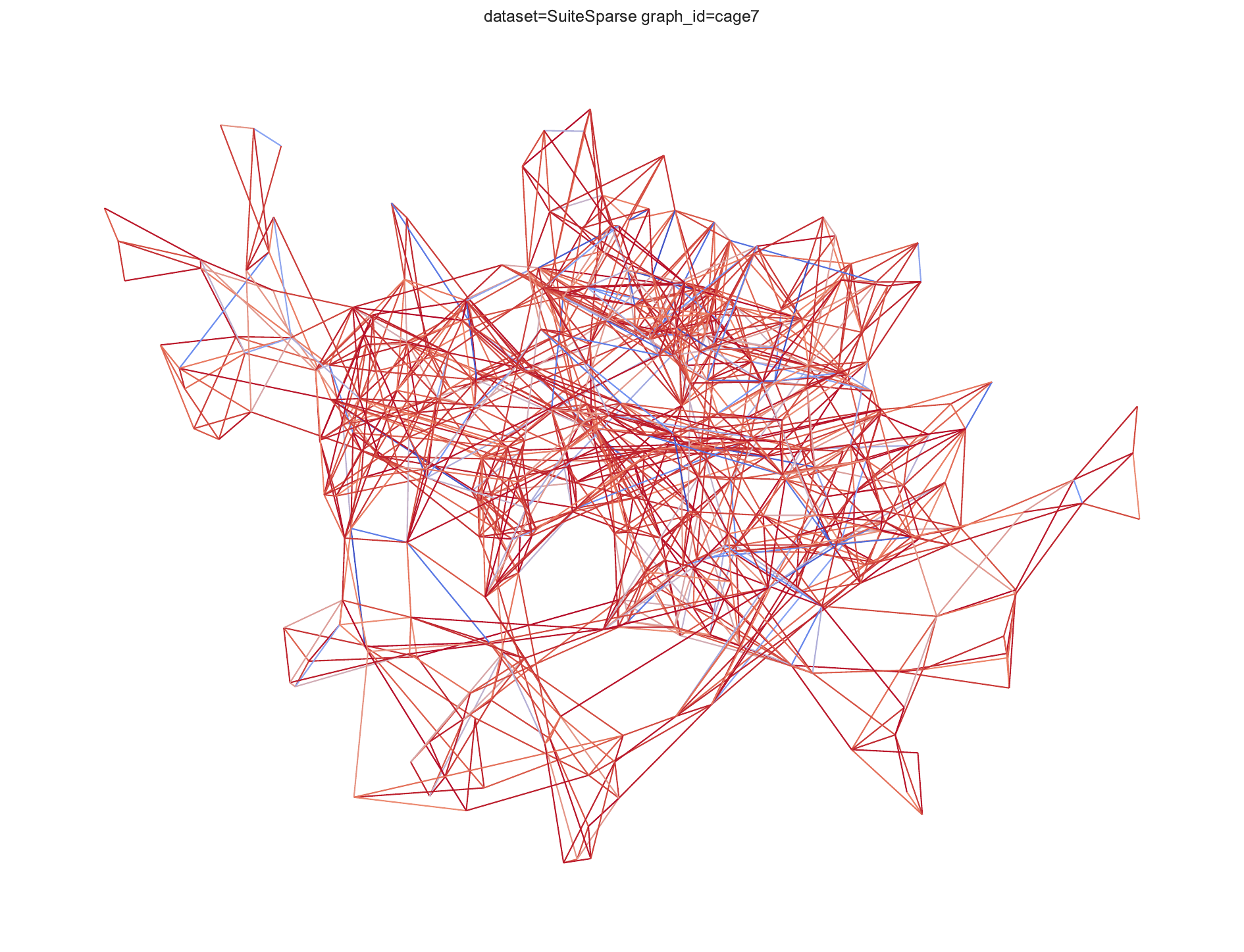} &
\imgcell{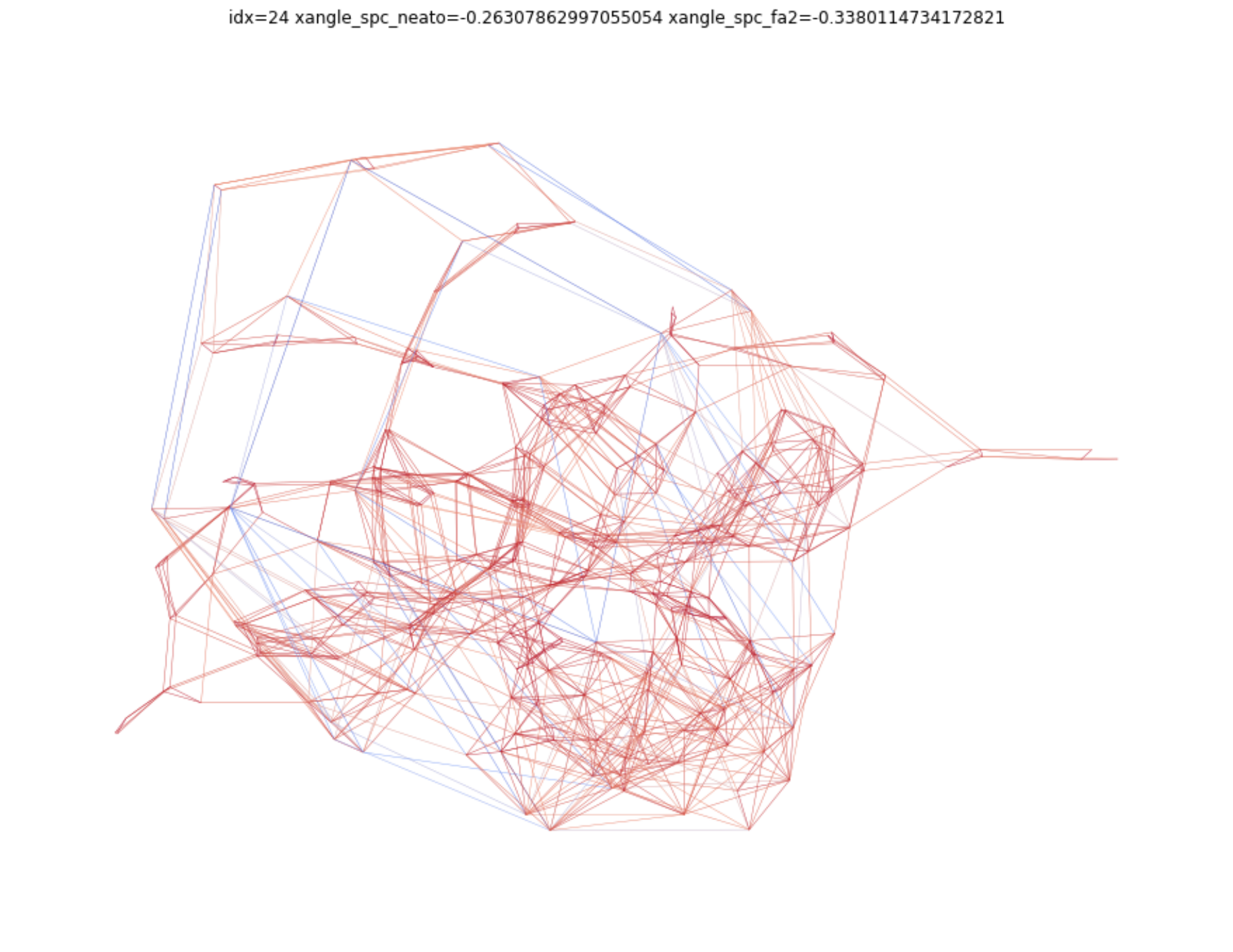} &
\imgcell{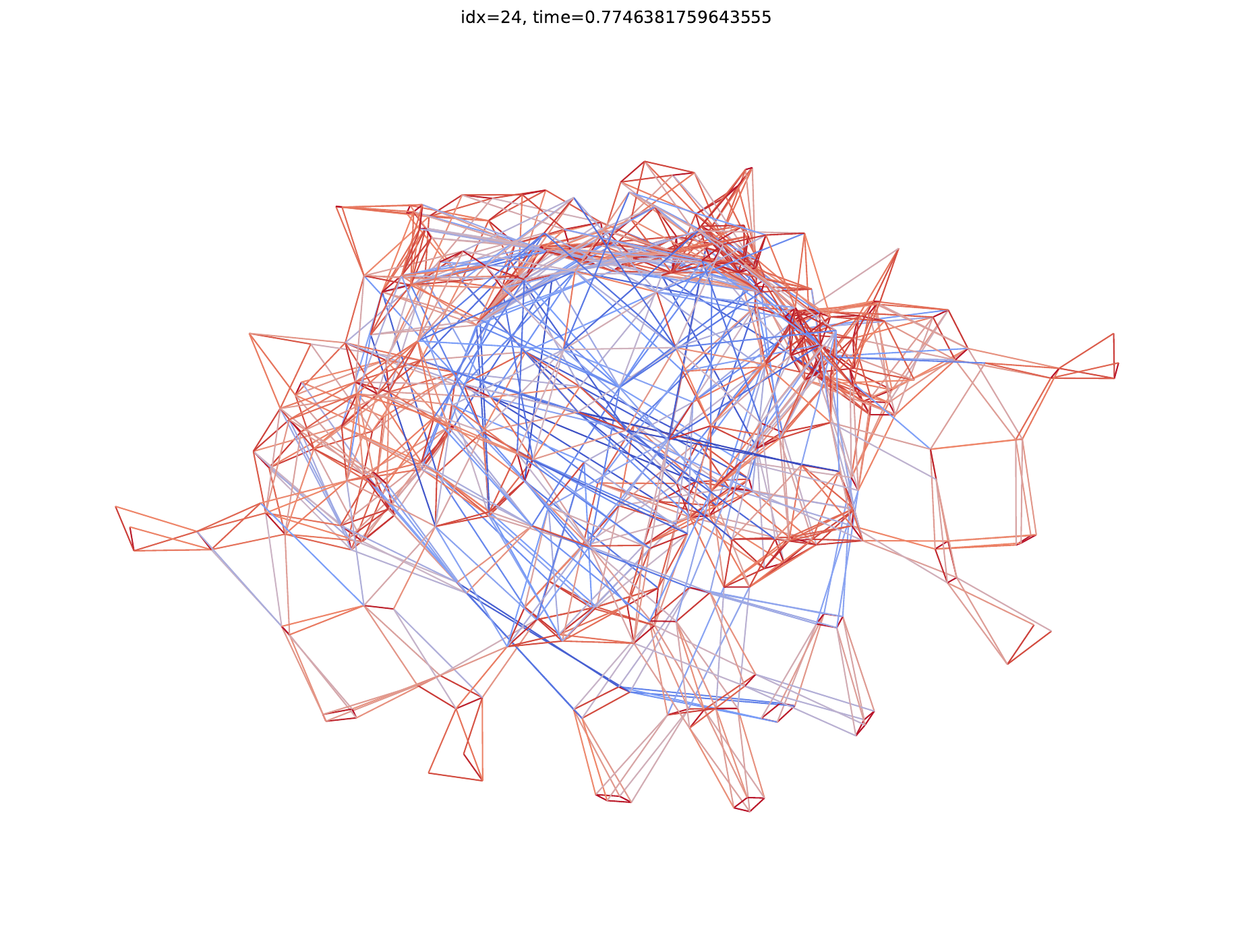} &
\imgcell{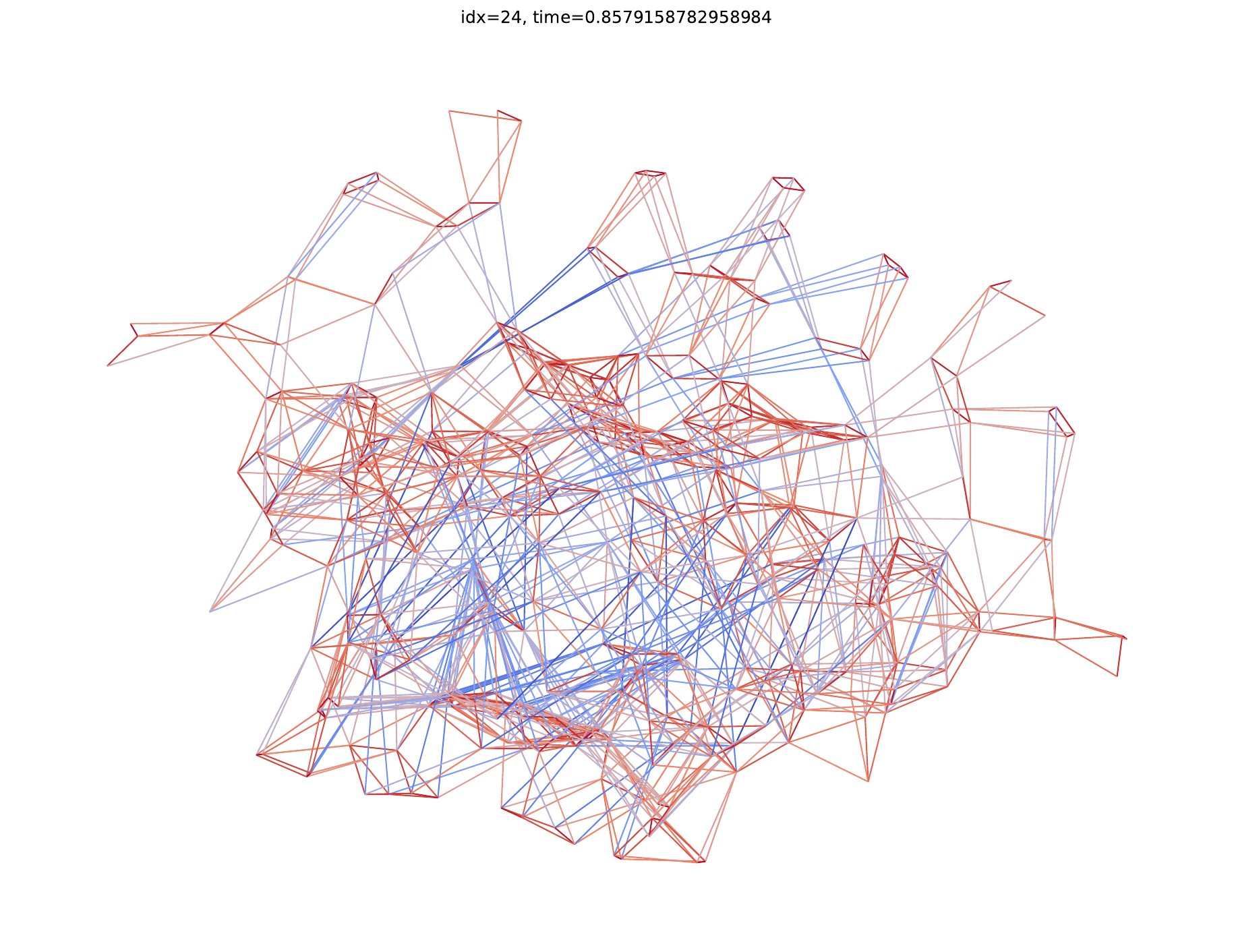} &
\imgcell{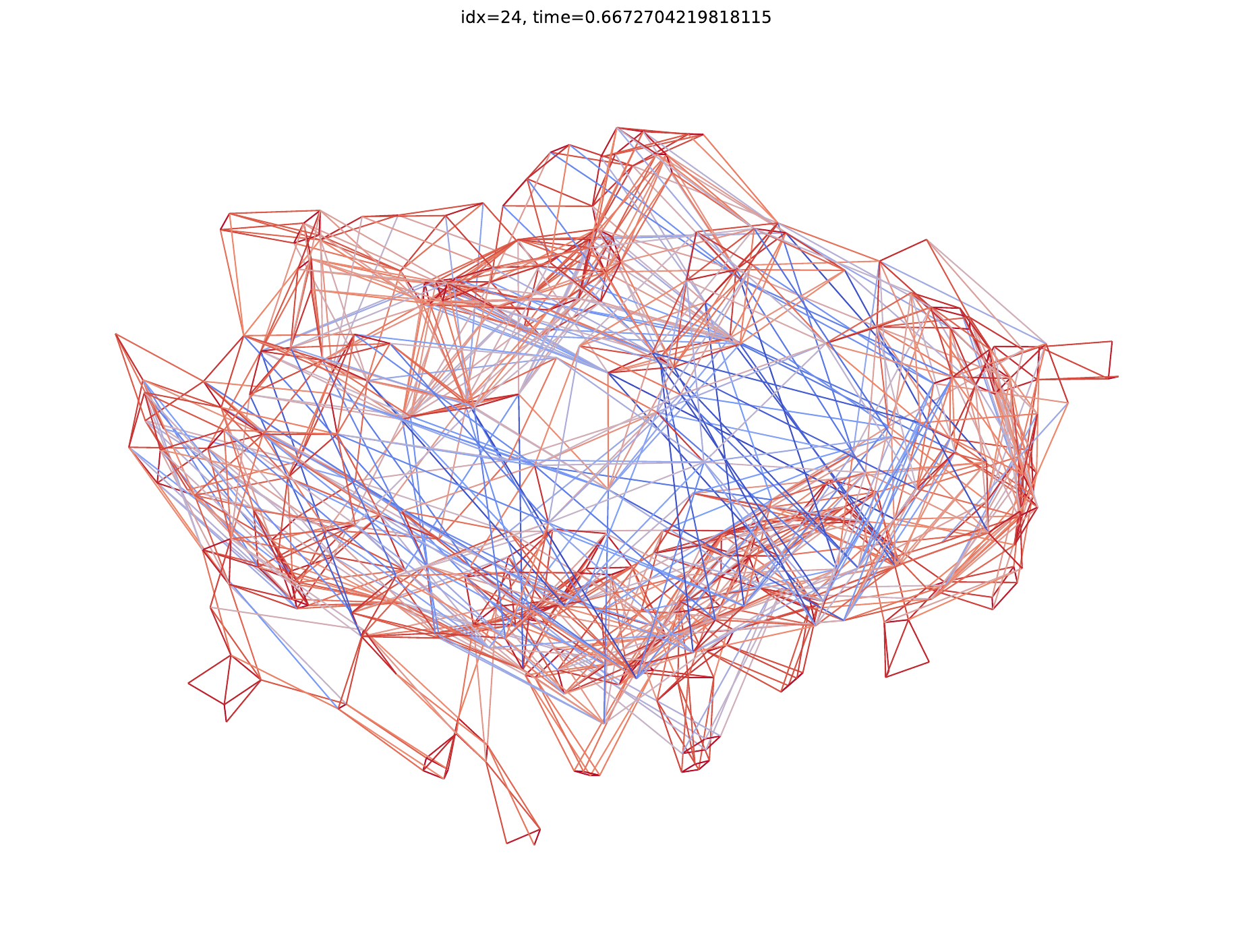} \\

&
t = 0.04s &
t = 6.96s &
t = 9.47s &
t = 0.55s &
t = 7200.00s &
t = 0.40s &
t = 0.32s &
t = 0.42s &
t = 0.46s &
t = 0.52s &
t = 0.39s &
t = 0.32s \\

\makecell{\bfseries gre\_343\\N = 343\\M = 1092} &
\imgcell{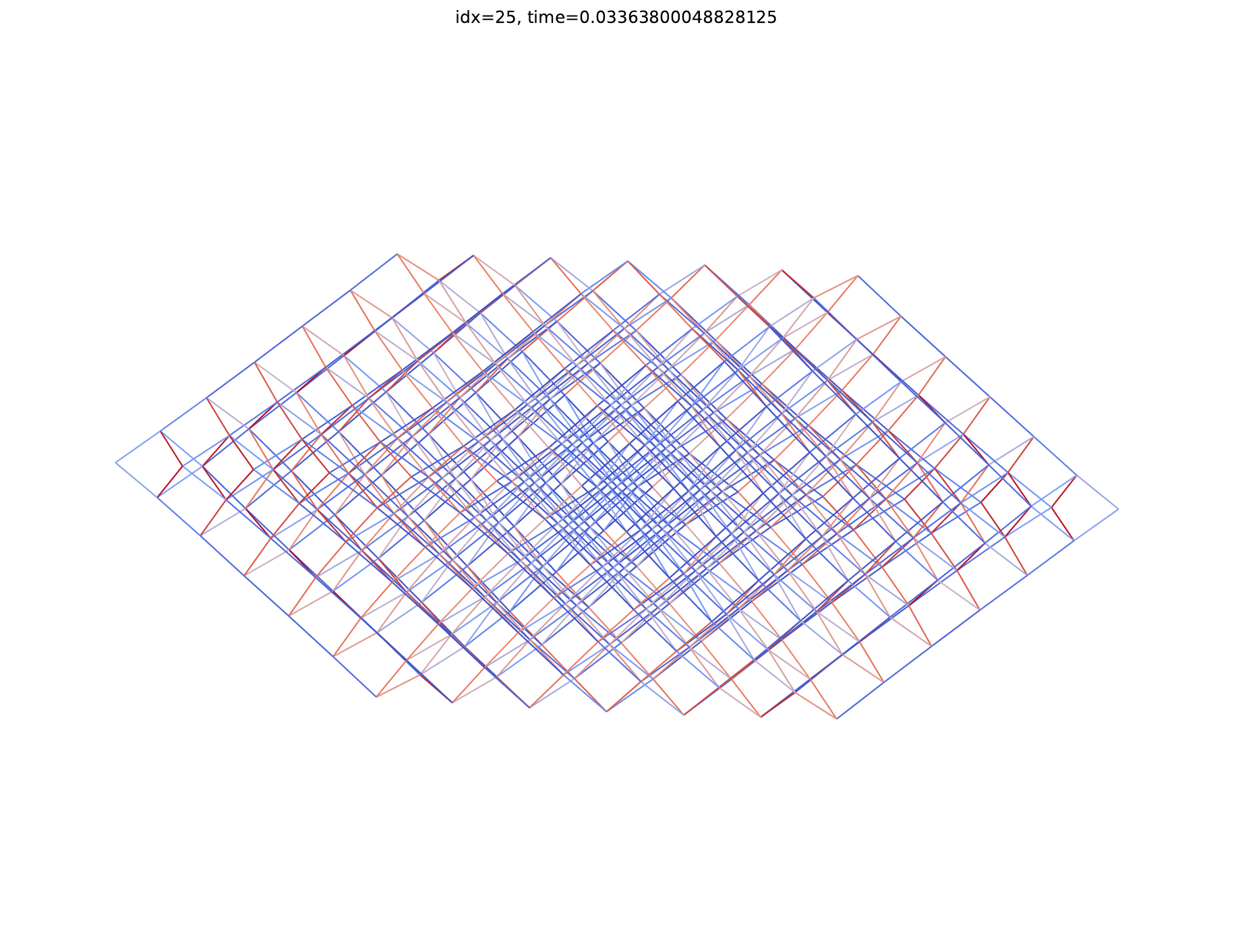} &
\imgcell{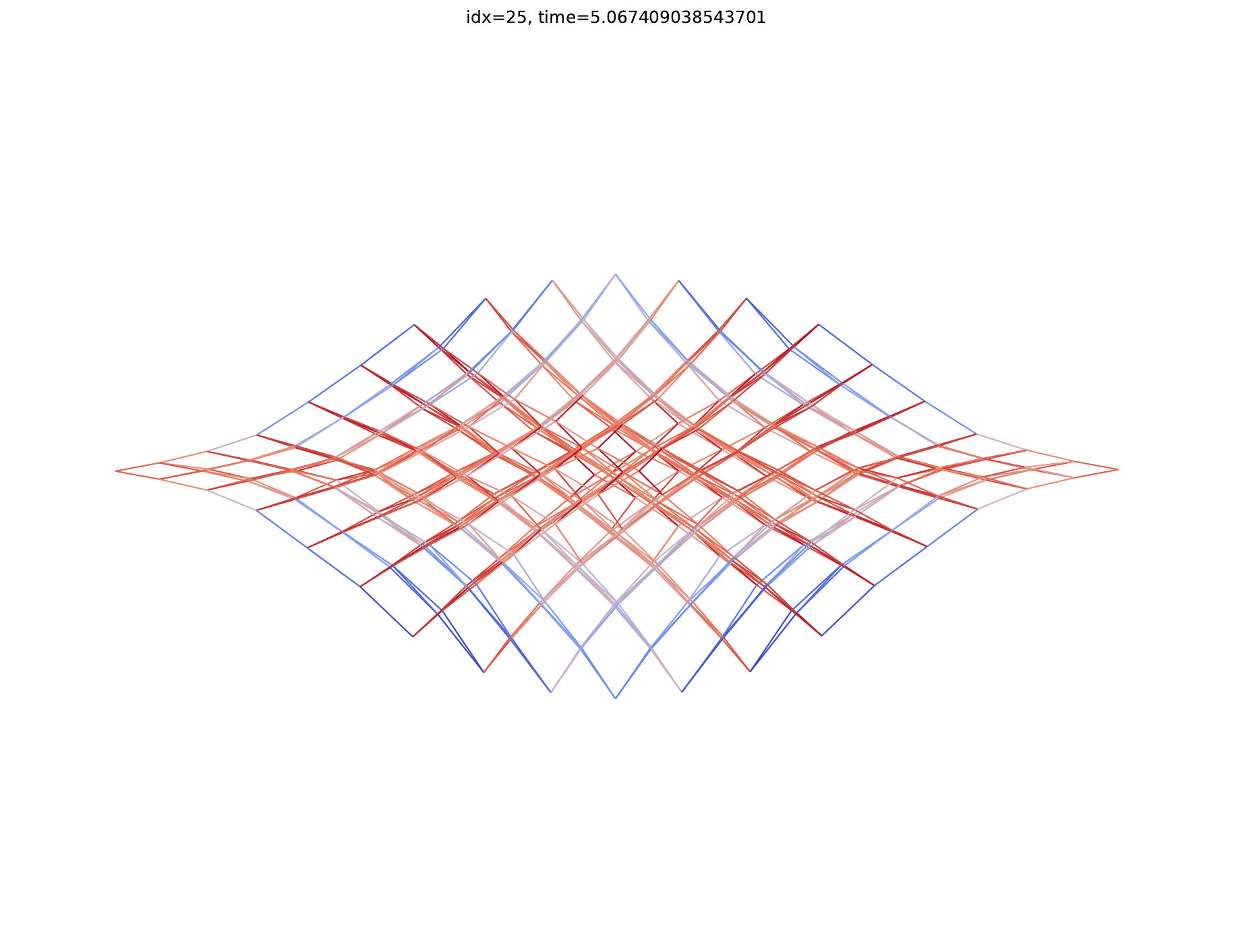} &
\imgcell{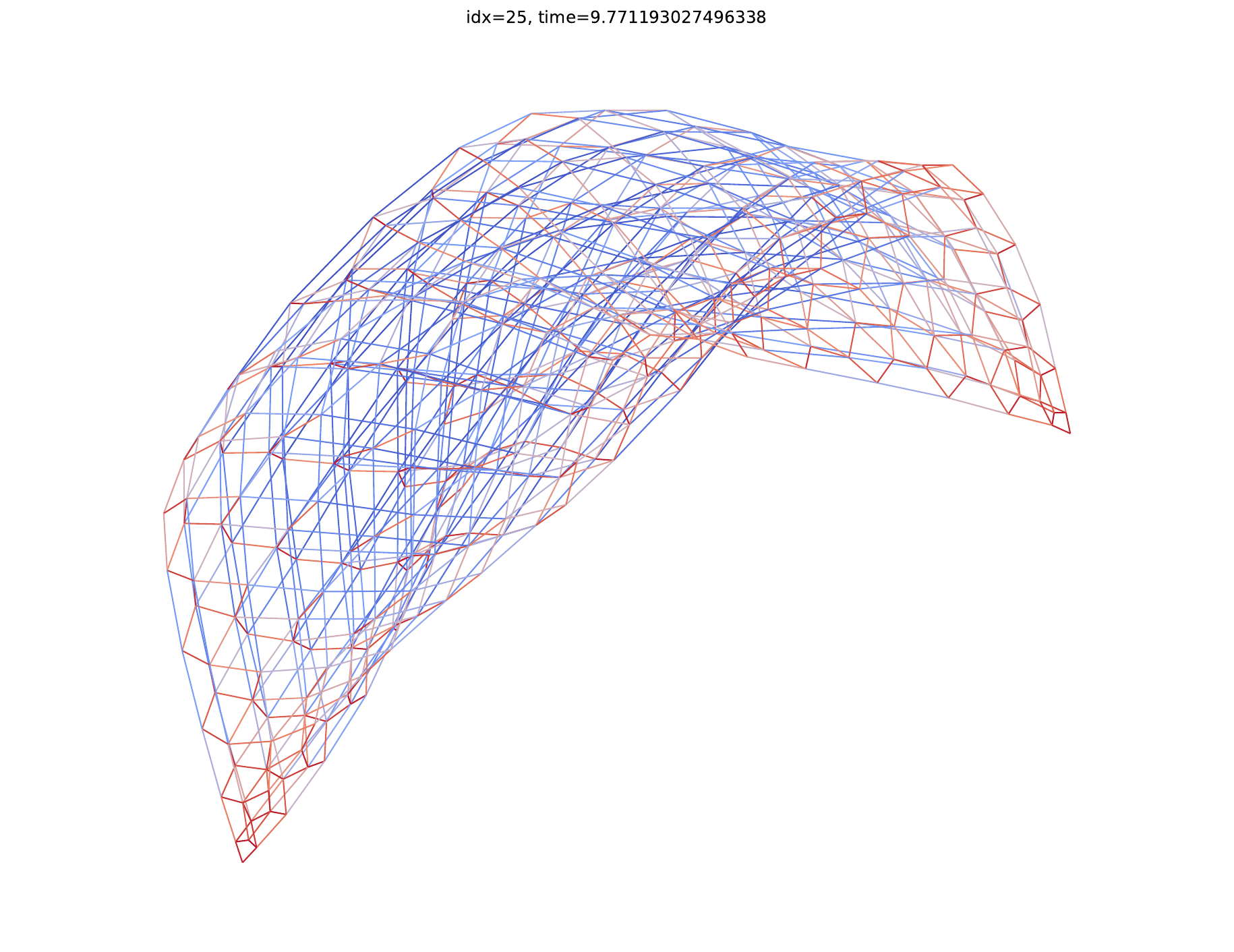} &
\imgcell{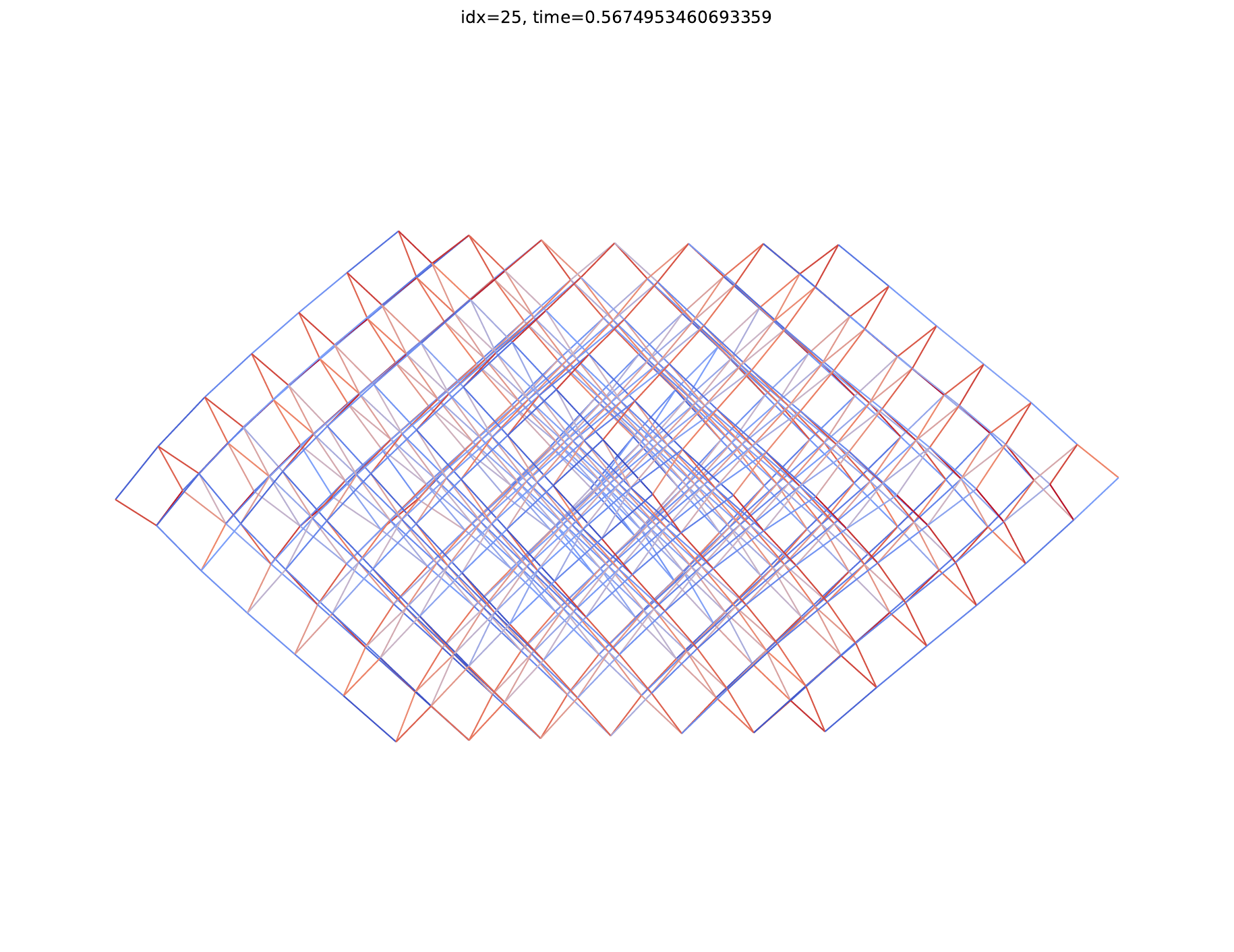} &
\imgcell{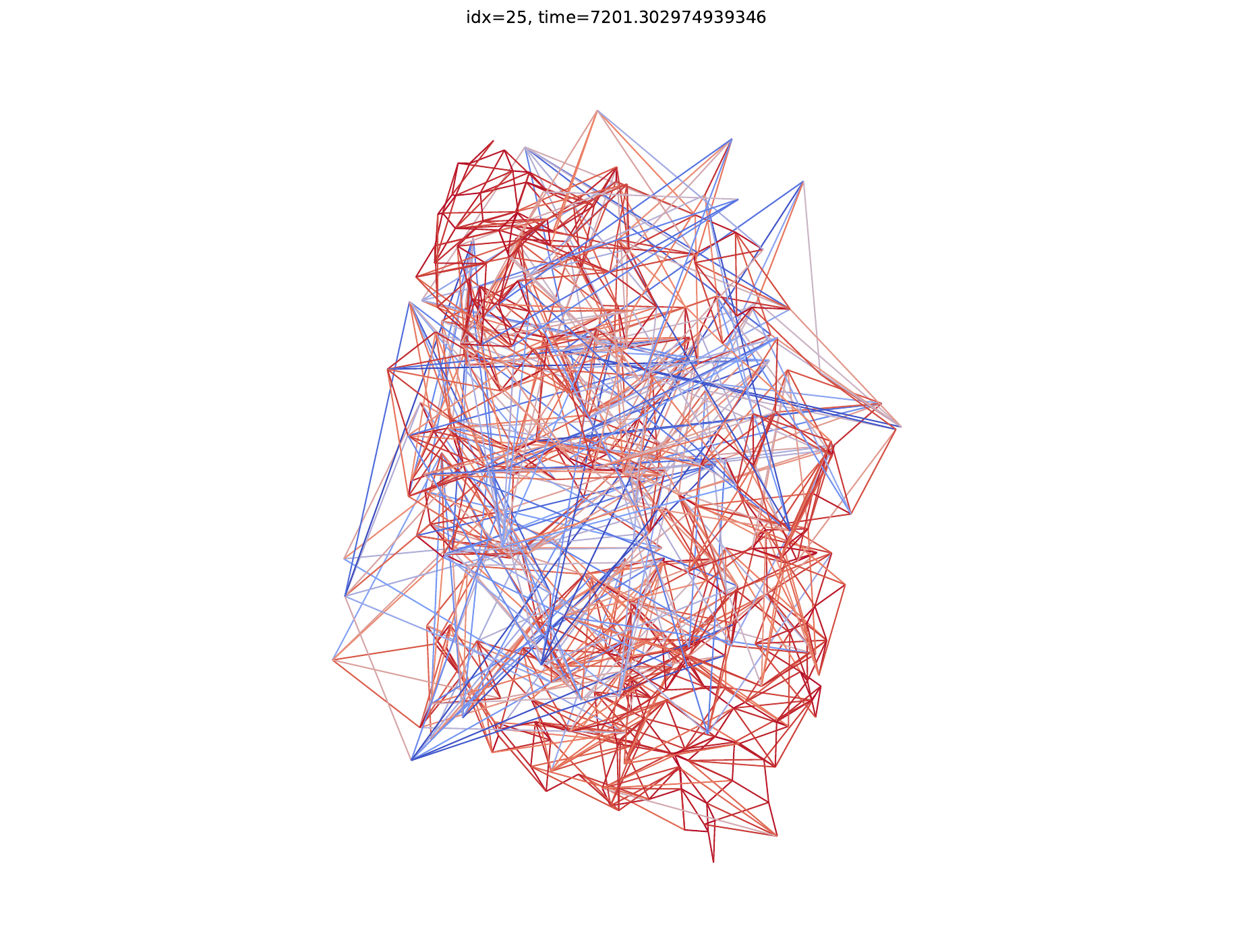} &
\imgcell{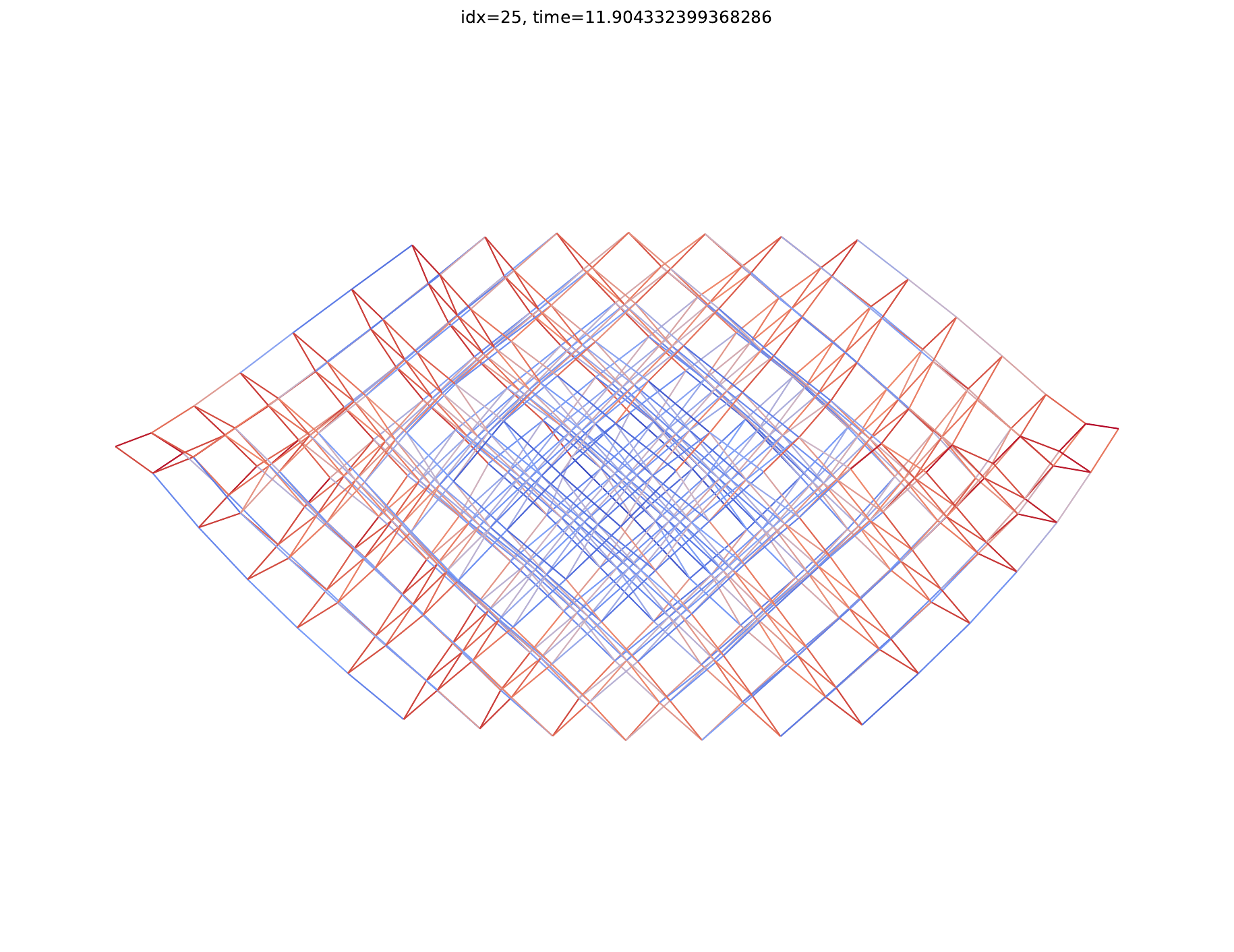} &
\imgcell{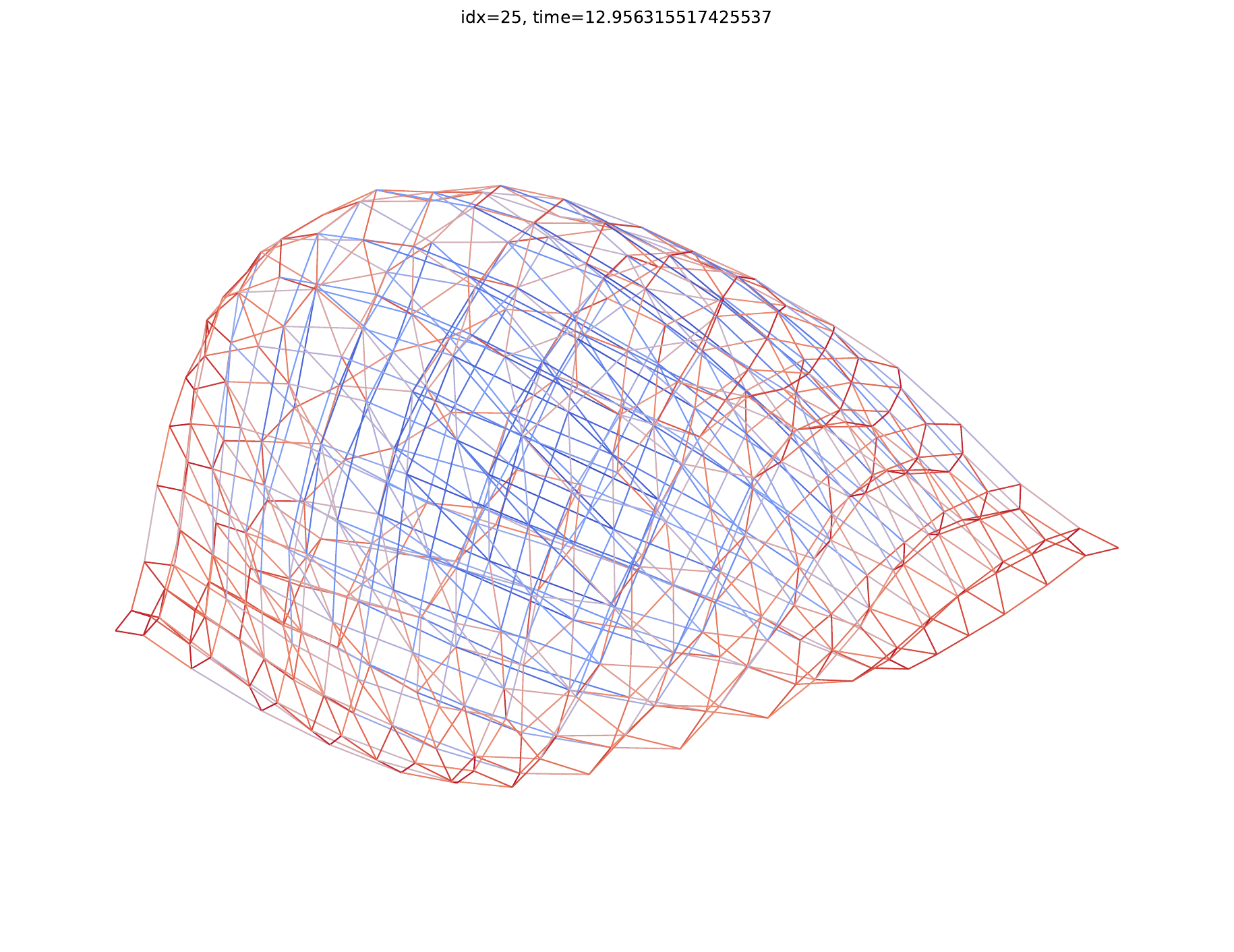} &
\imgcell{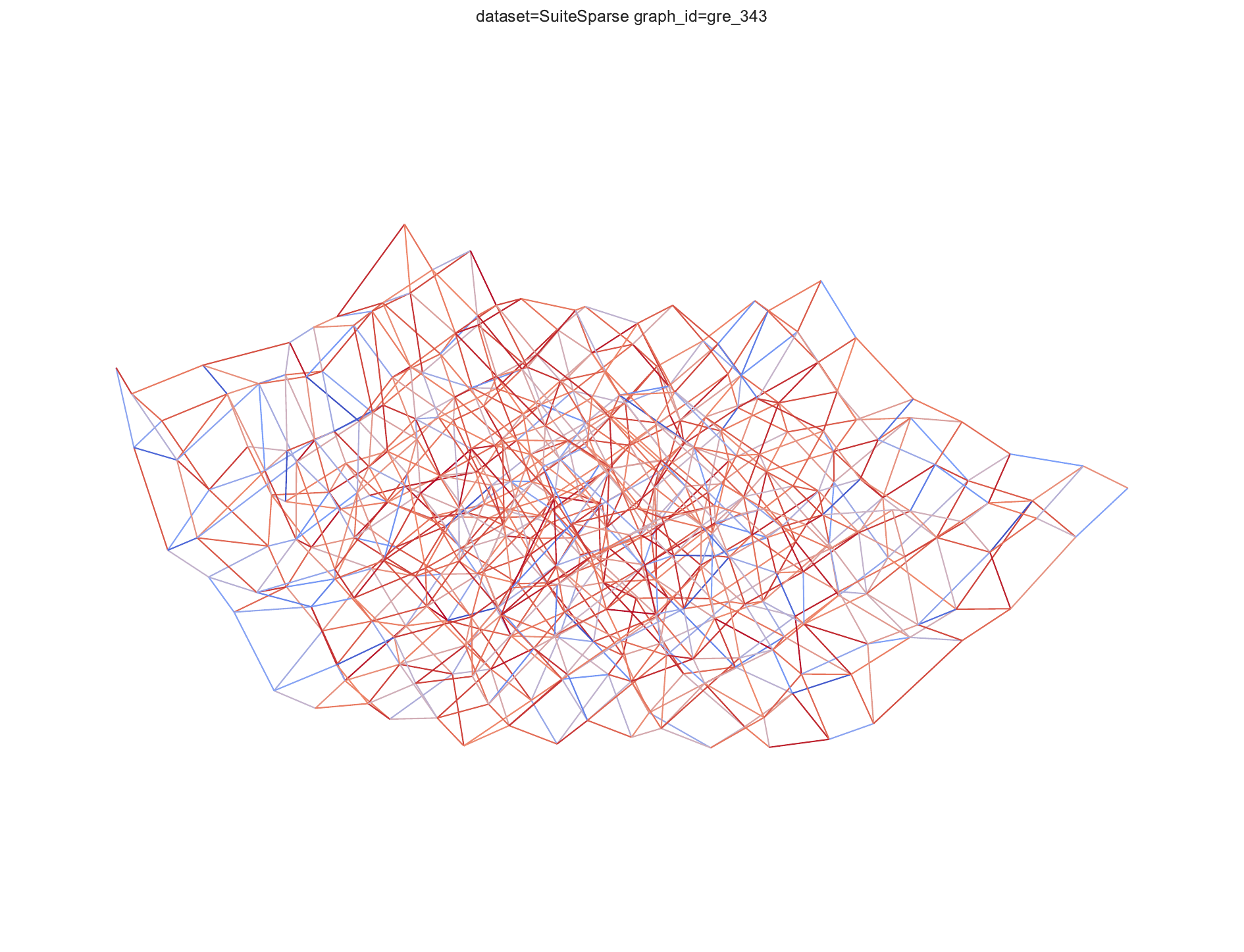} &
\imgcell{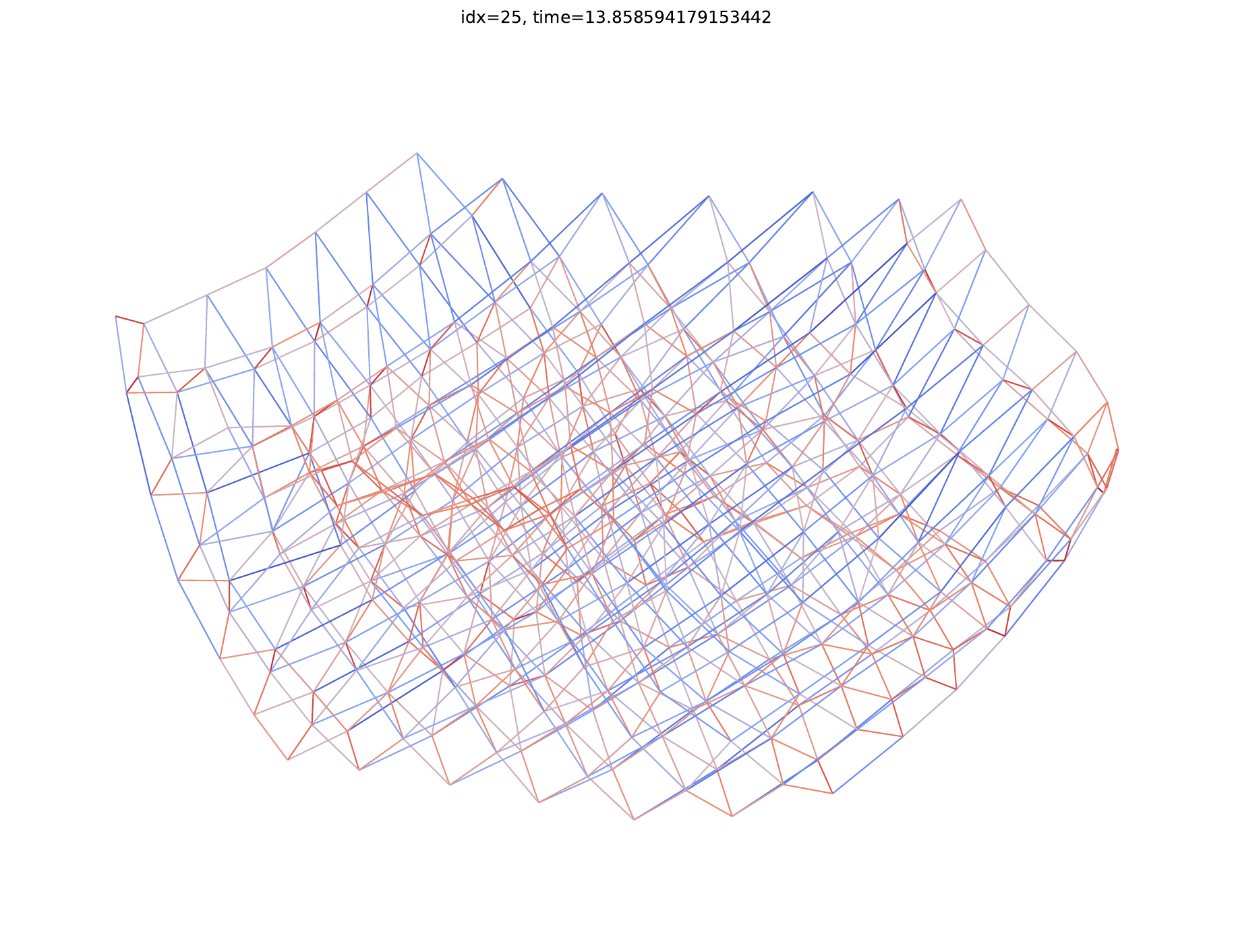} &
\imgcell{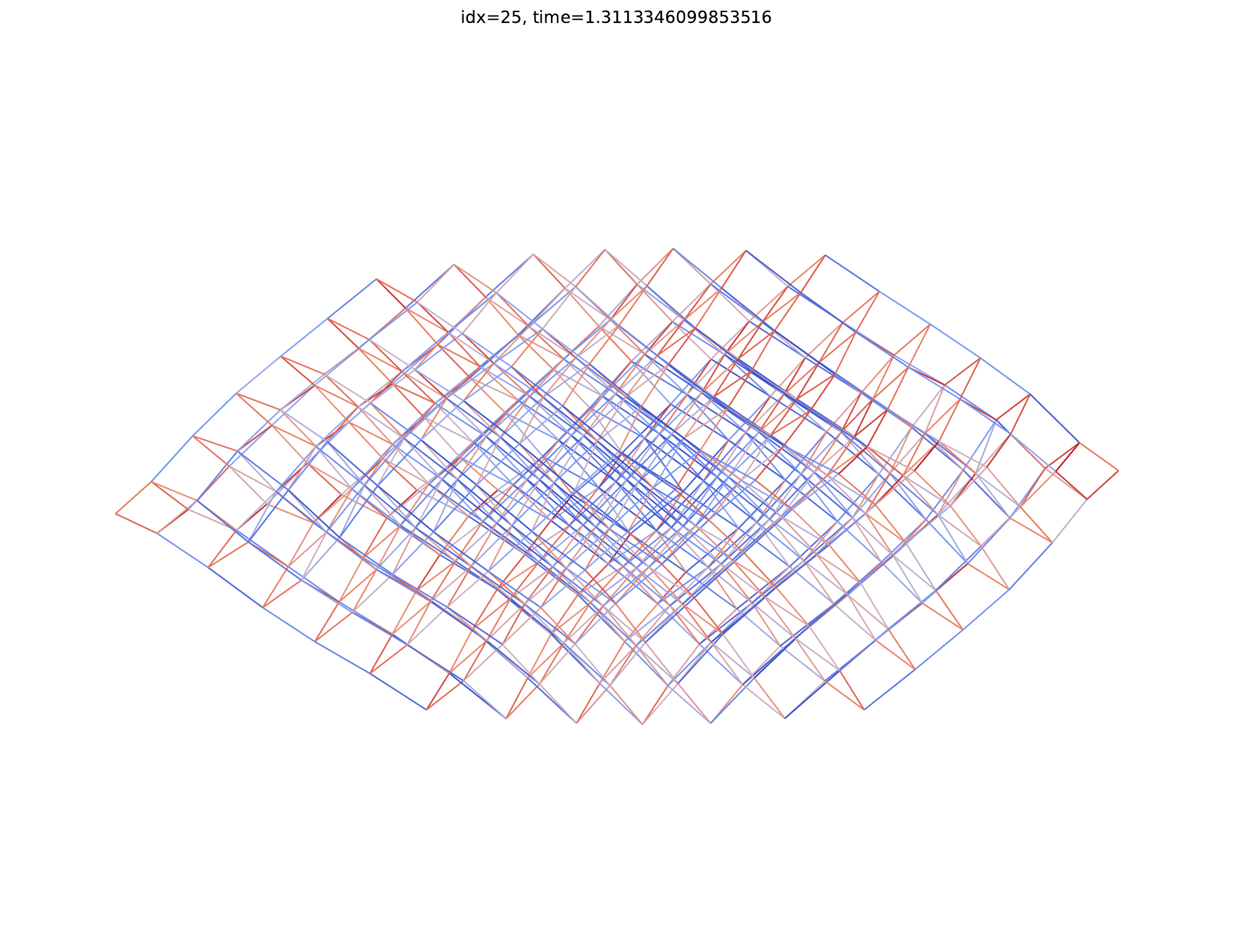} &
\imgcell{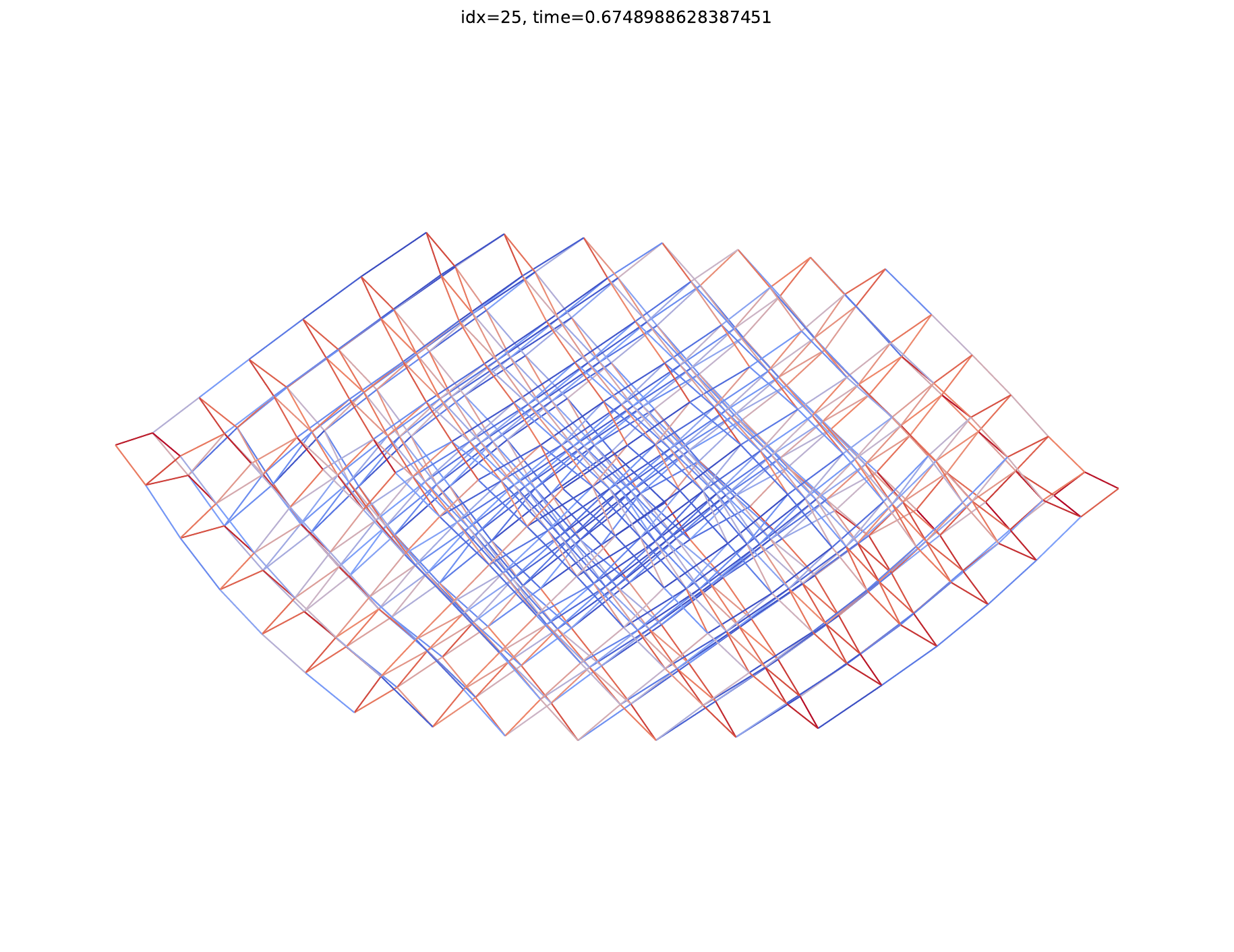} &
\imgcell{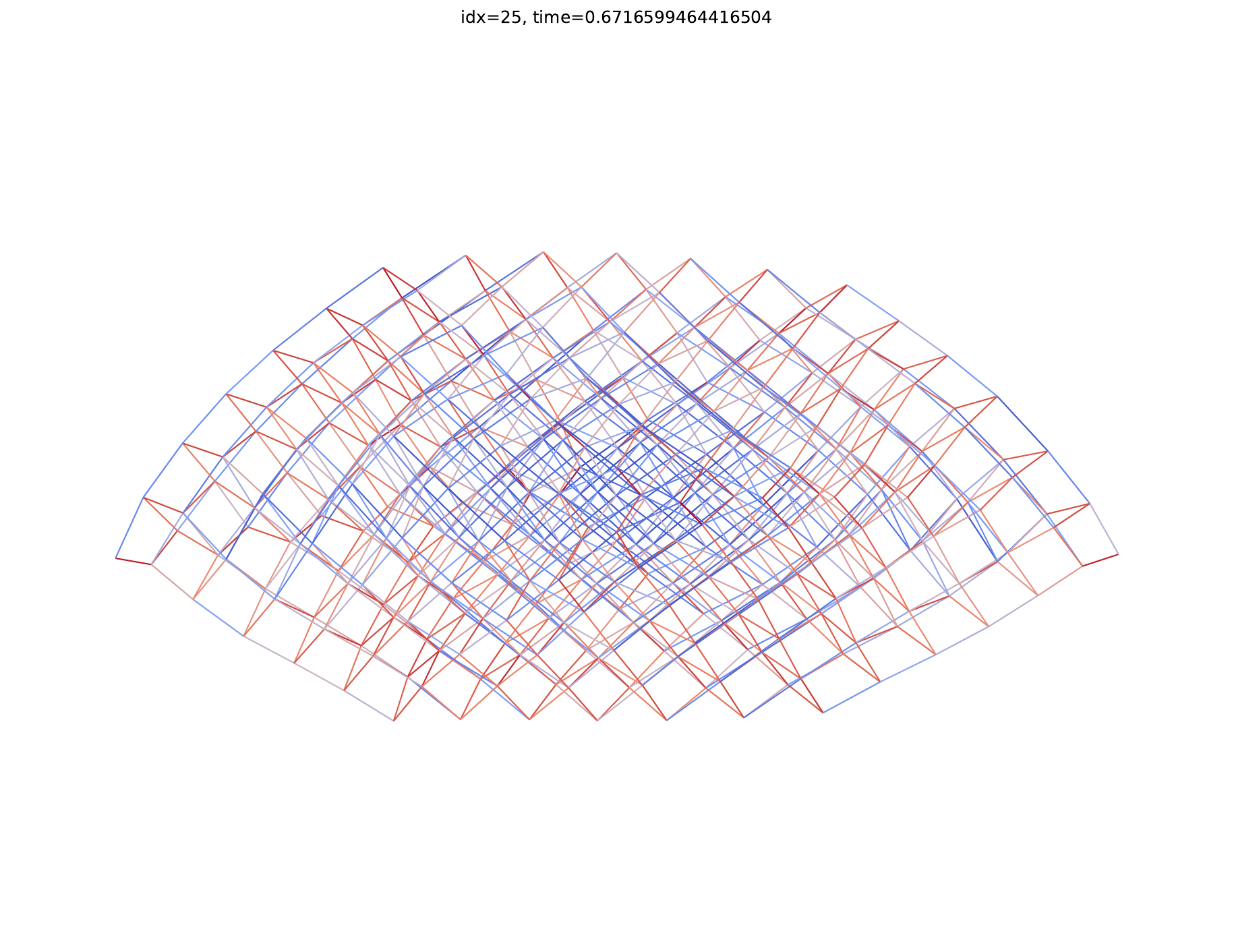} \\

&
t = 0.03s &
t = 5.07s &
t = 9.77s &
t = 0.57s &
t = 7200.00s &
t = 0.52s &
t = 0.46s &
t = 0.52s &
t = 0.61s &
t = 0.43s &
t = 0.33s &
t = 0.49s \\

\makecell{\bfseries dwt\_361\\N = 361\\M = 1296} &
\imgcell{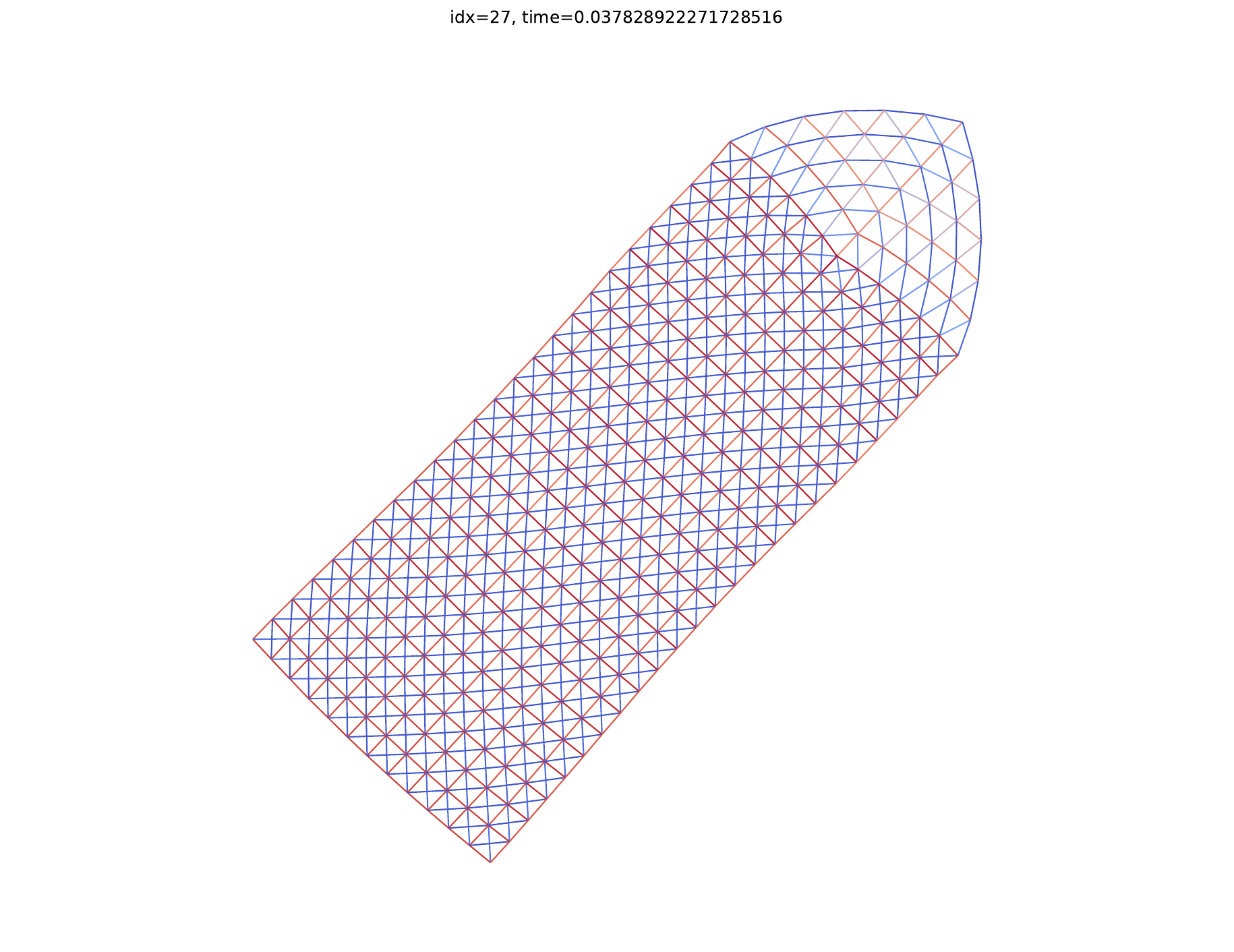} &
\imgcell{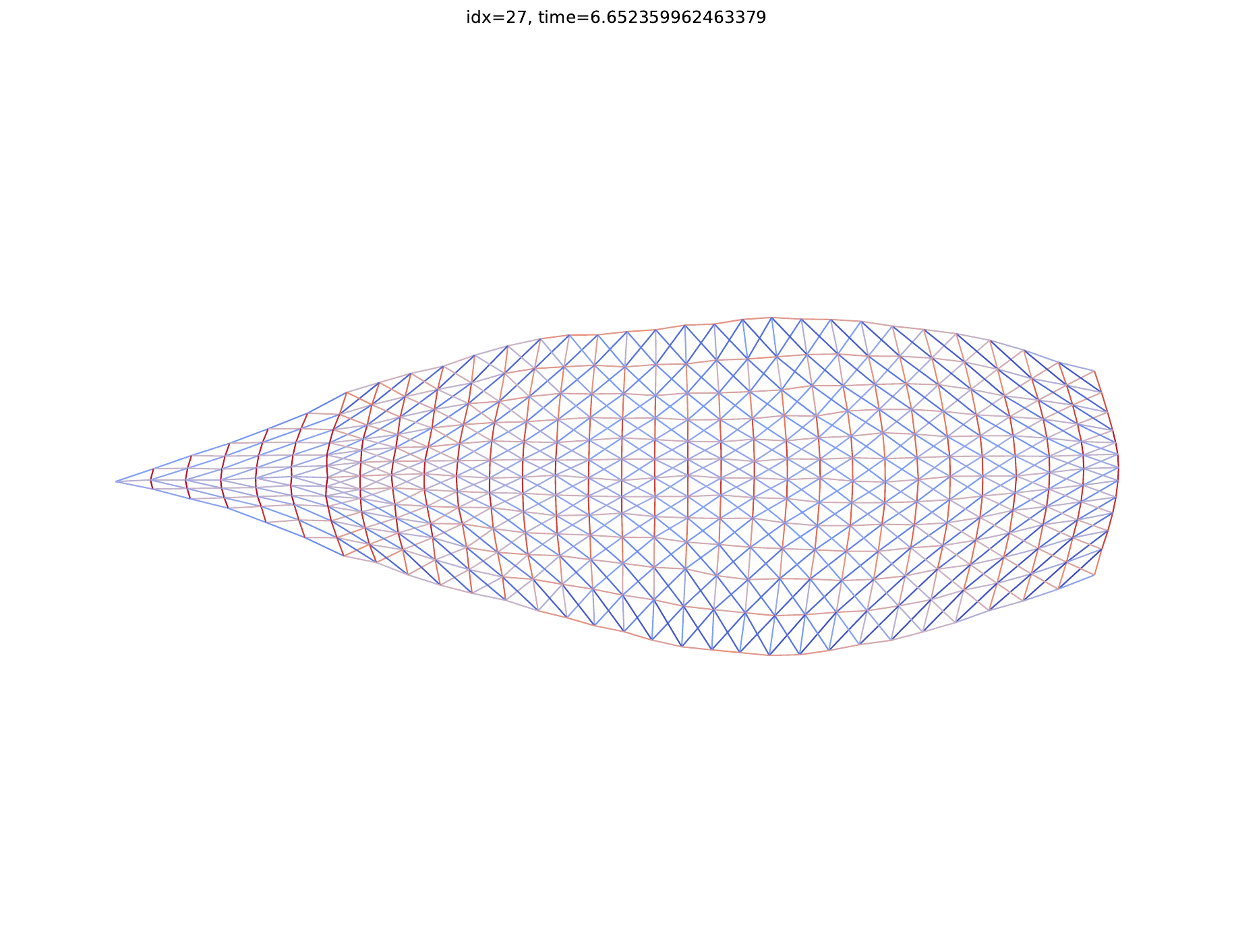} &
\imgcell{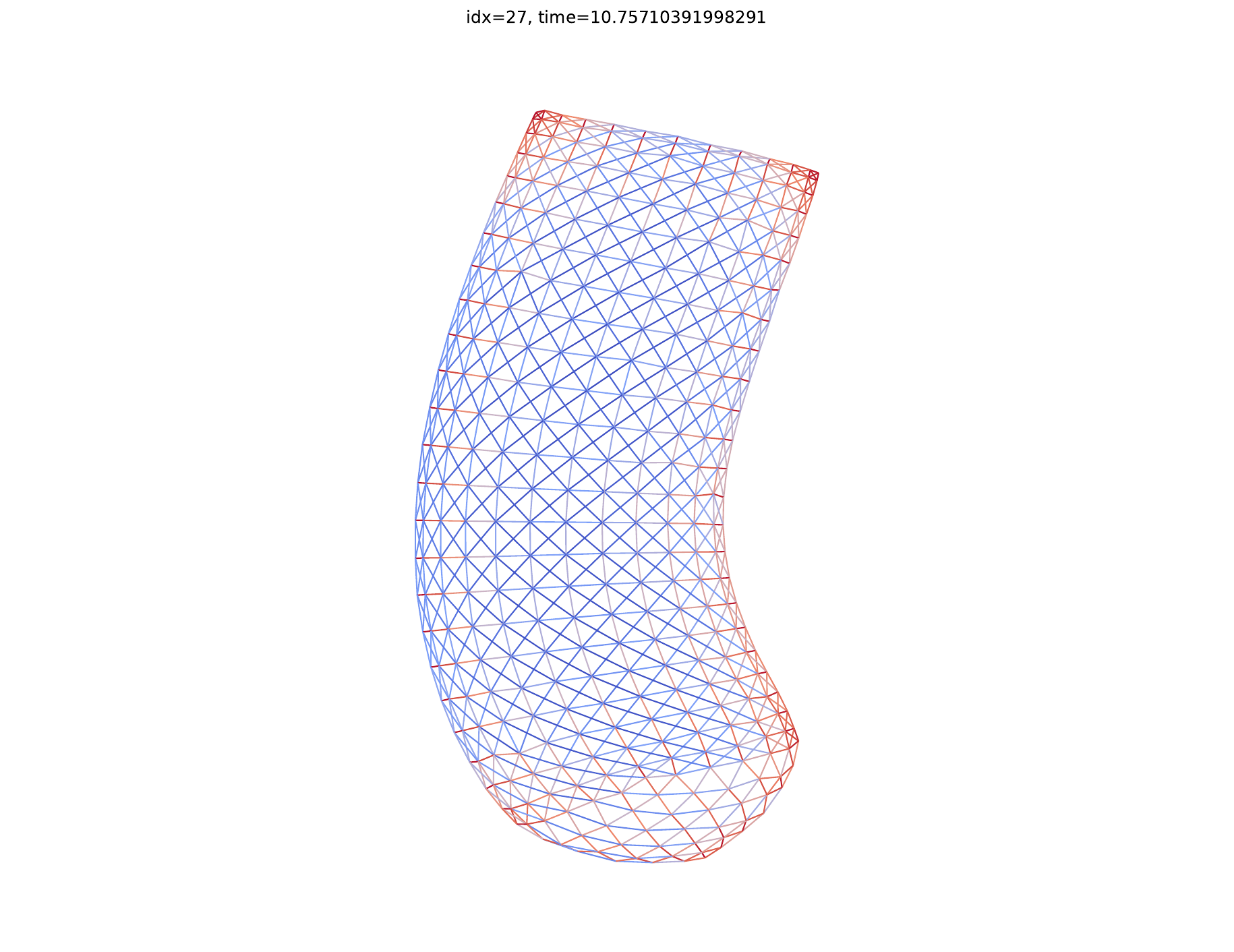} &
\imgcell{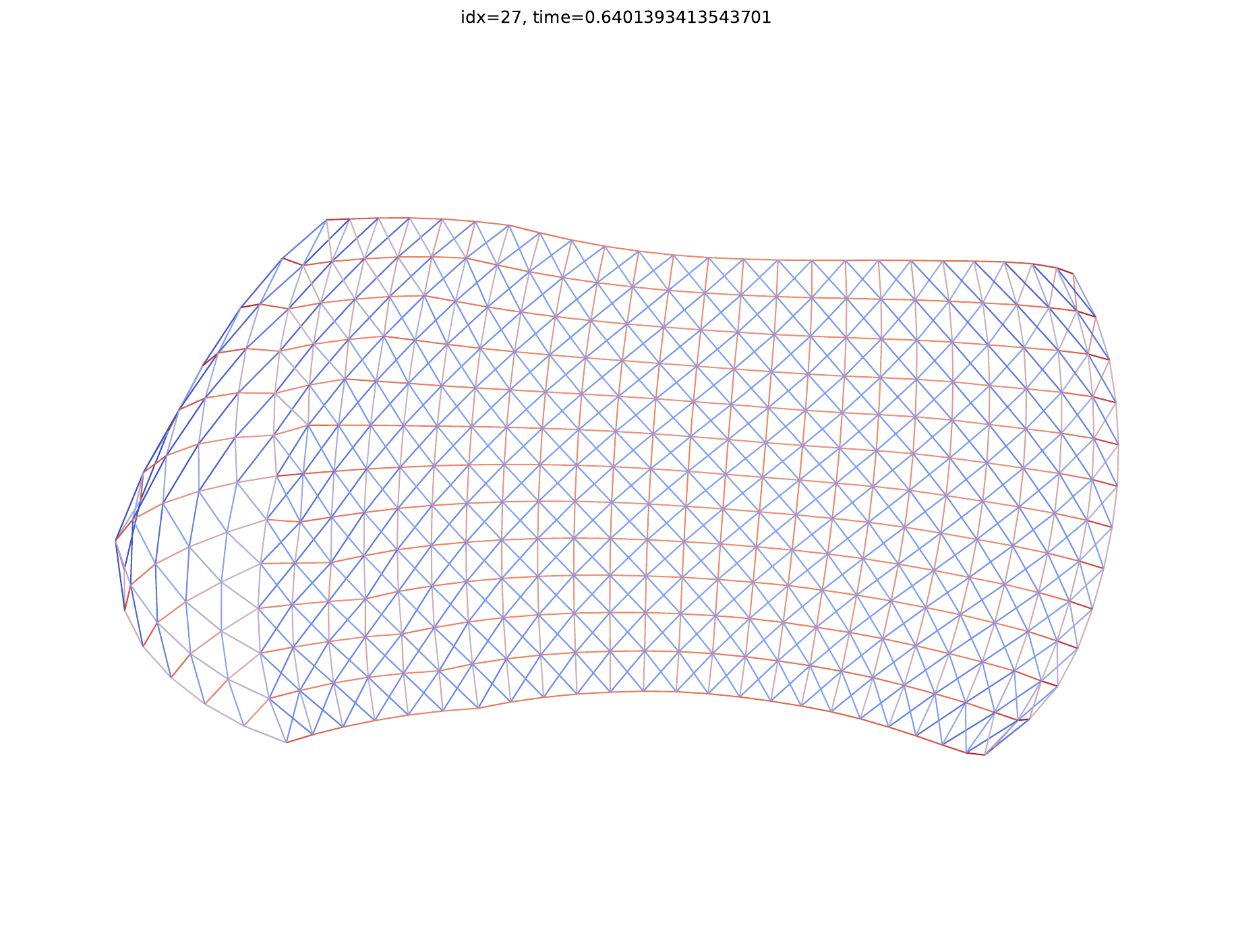} &
\imgcell{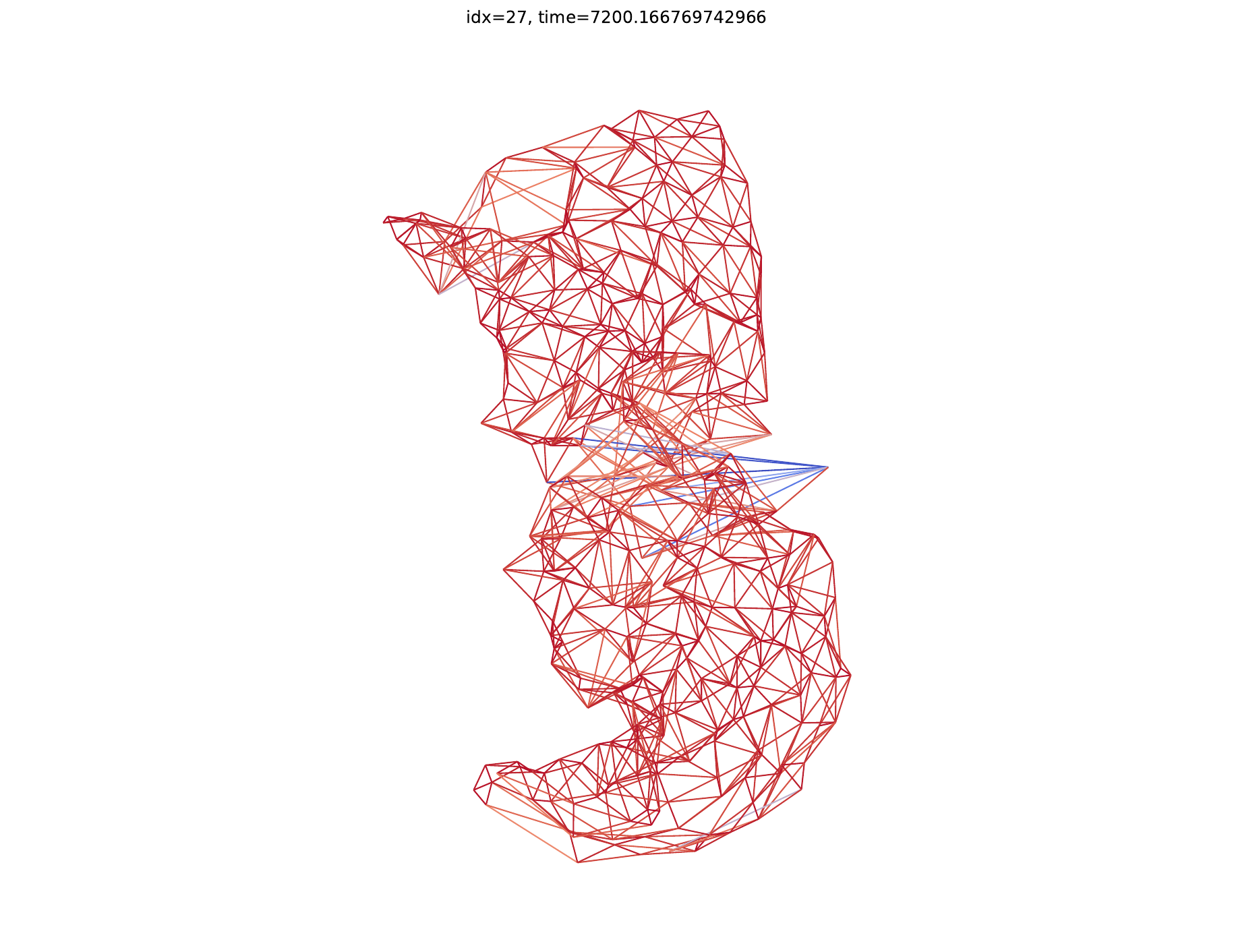} &
\imgcell{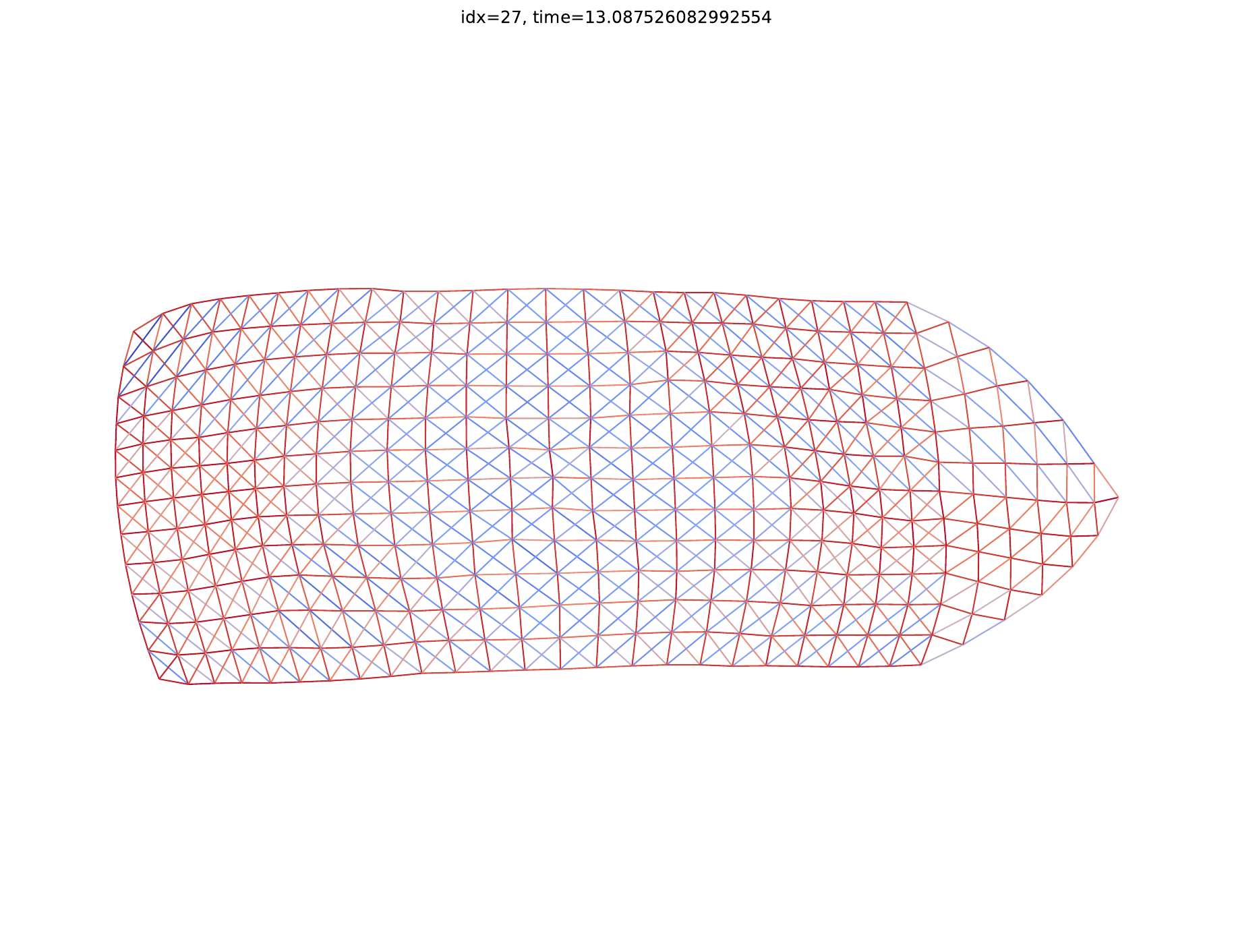} &
\imgcell{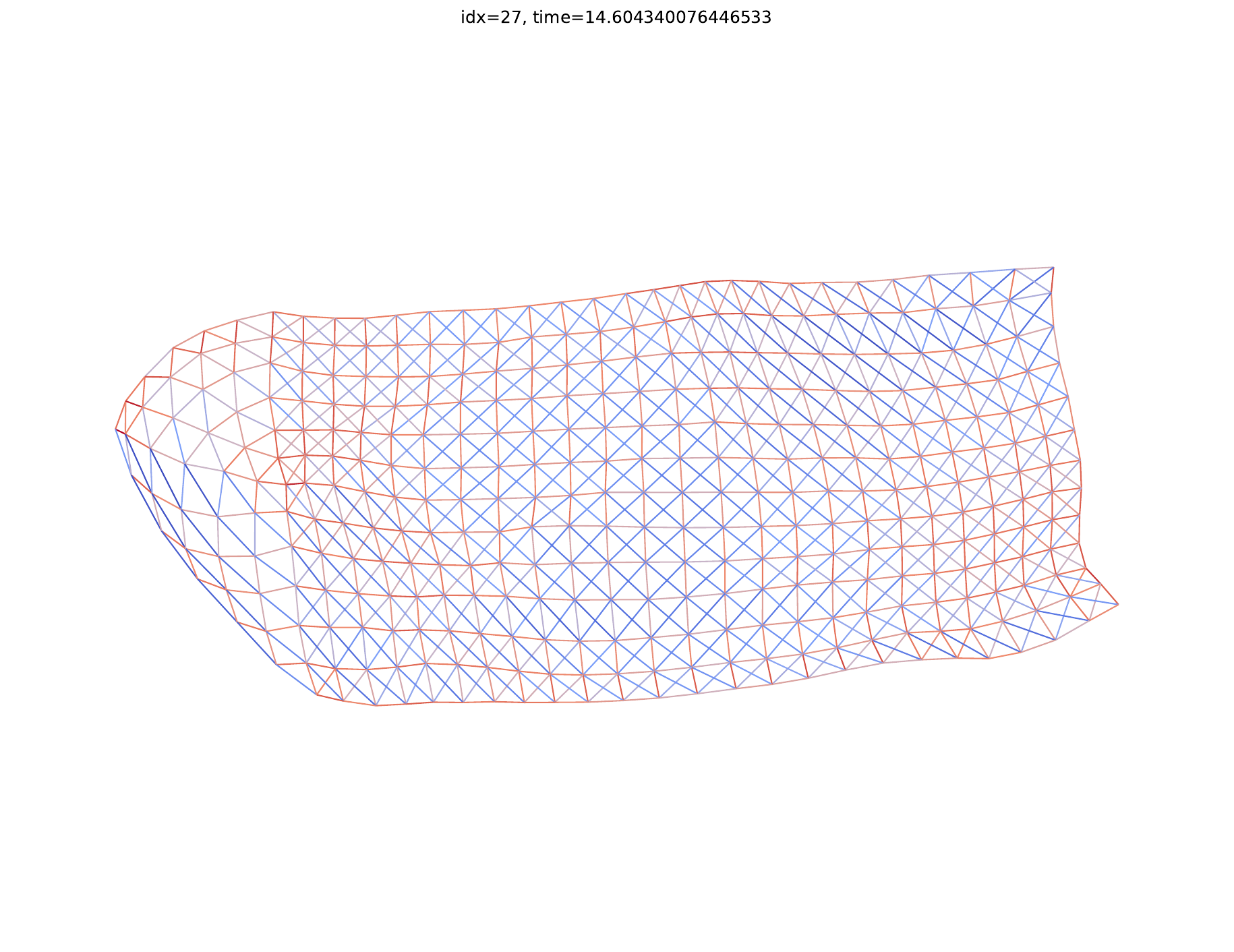} &
\imgcell{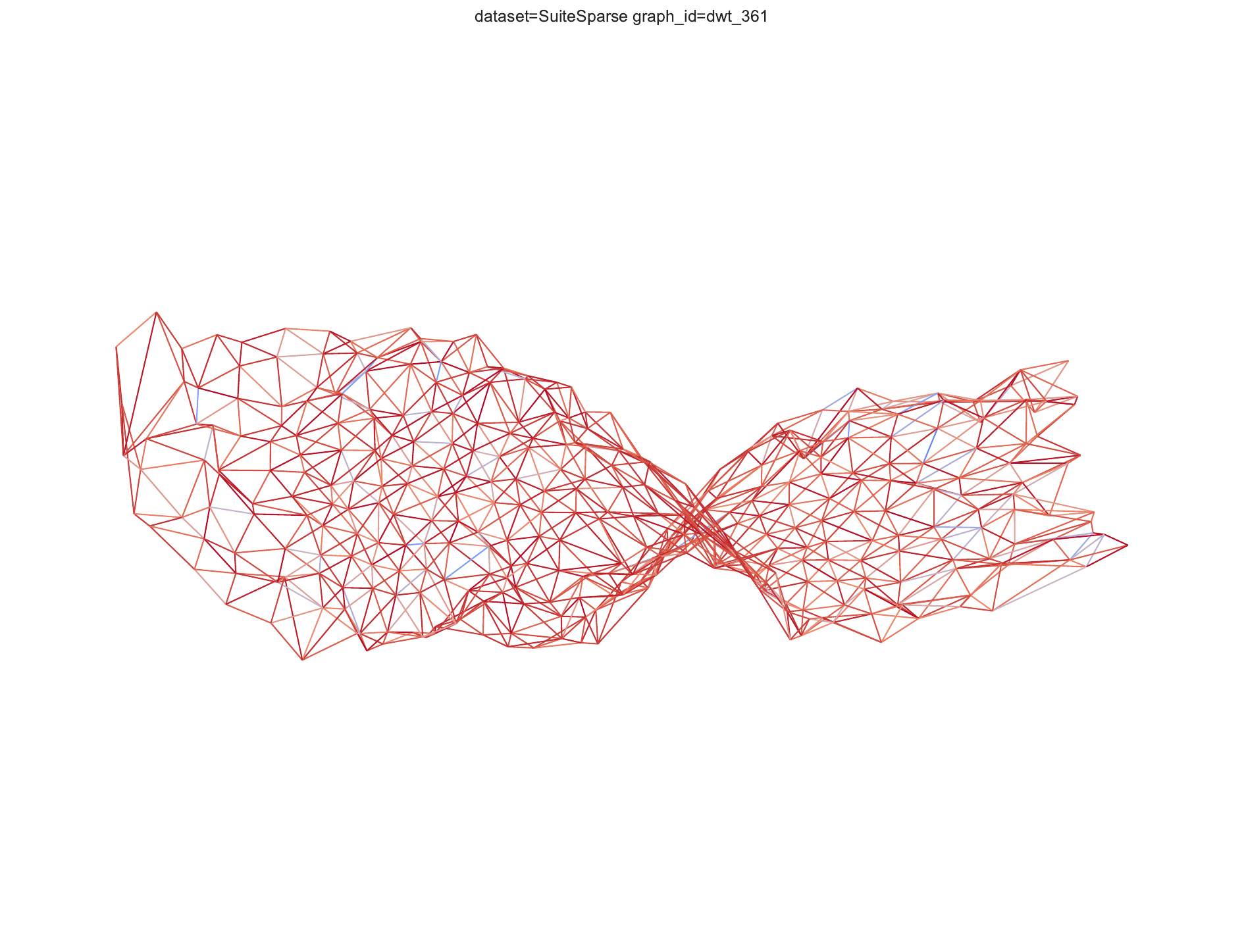} &
\imgcell{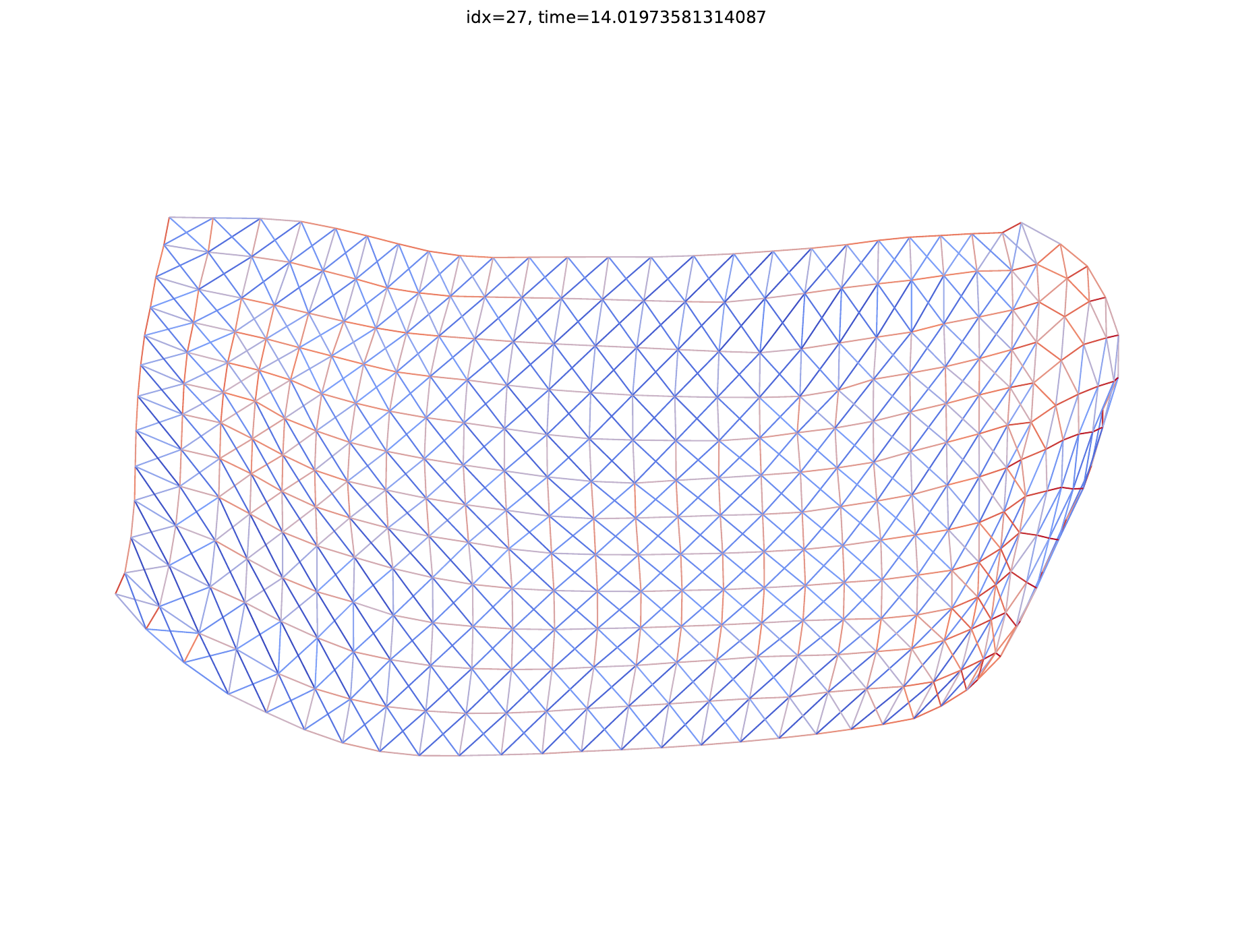} &
\imgcell{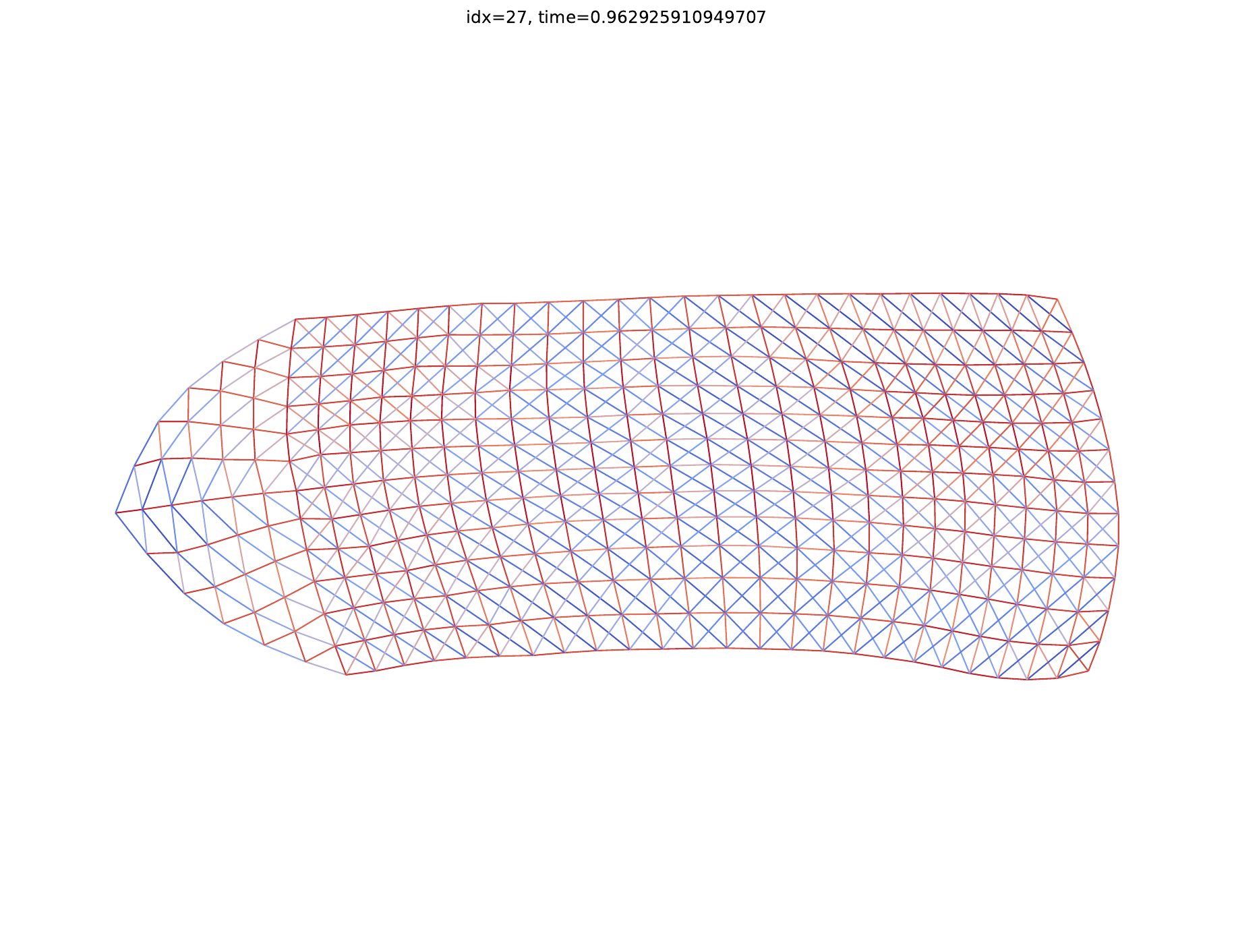} &
\imgcell{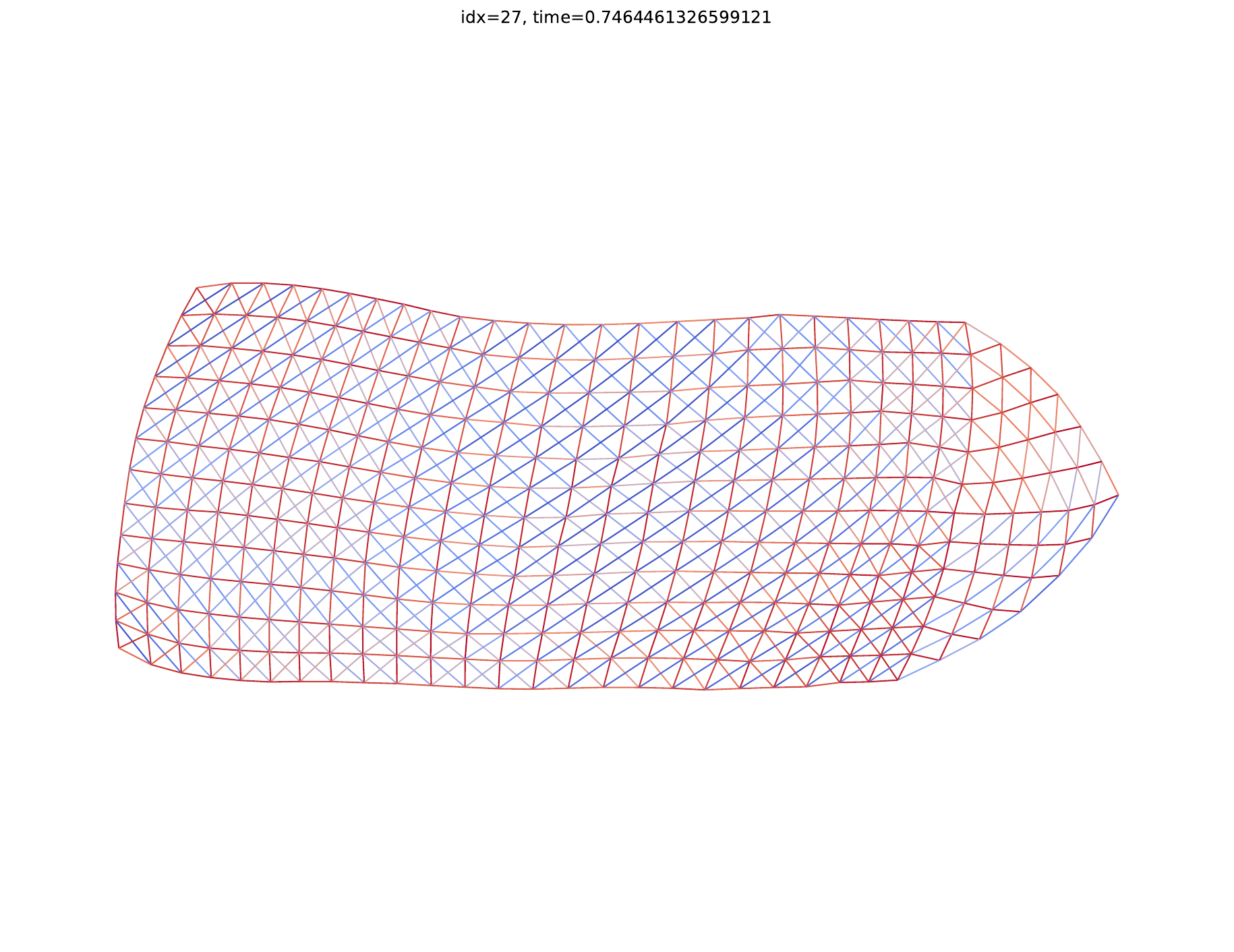} &
\imgcell{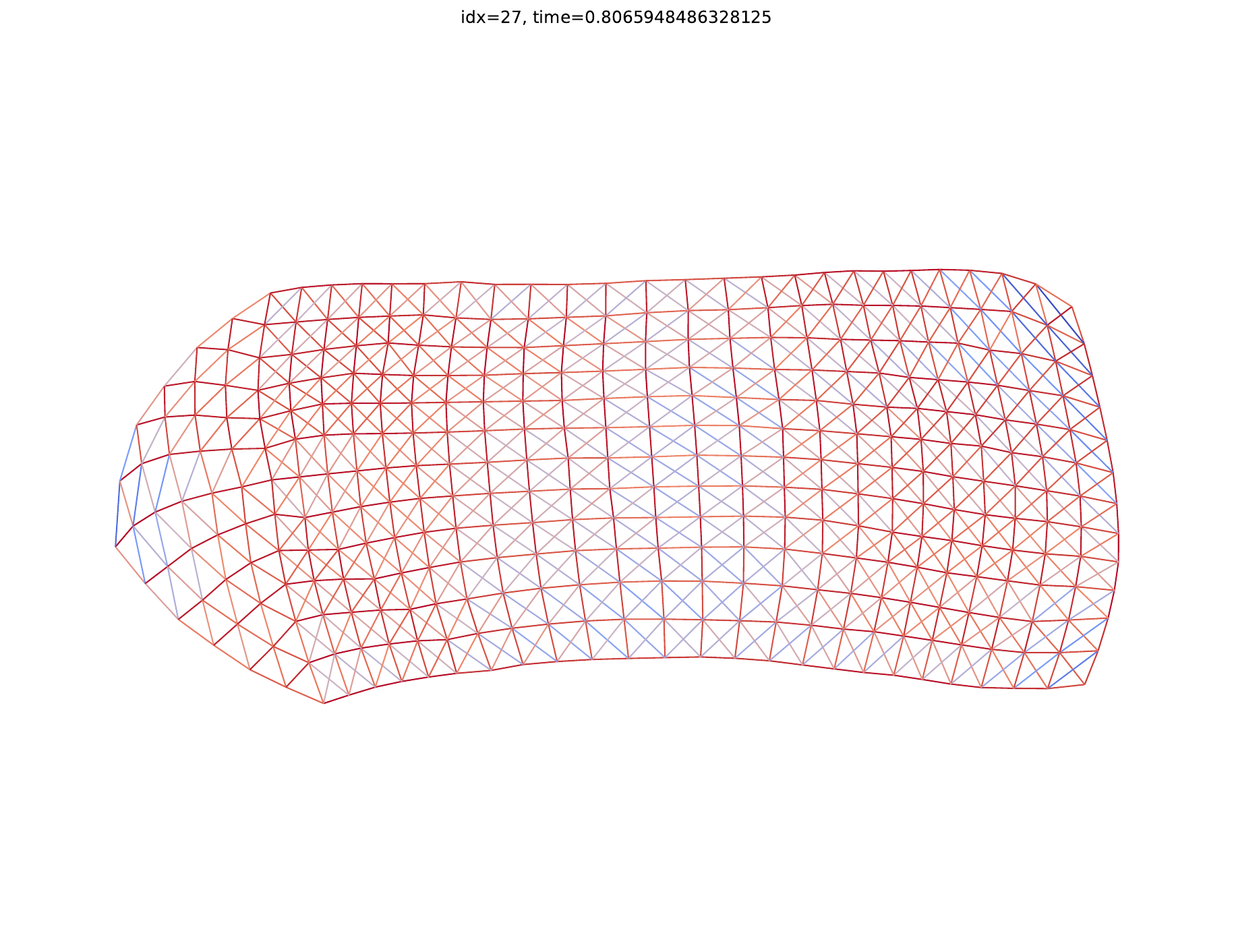} \\

&
t = 0.04s &
t = 6.65s &
t = 10.76s &
t = 0.64s &
t = 7200.00s &
t = 0.56s &
t = 0.61s &
t = 0.65s &
t = 0.73s &
t = 0.67s &
t = 0.69s &
t = 0.51s \\

\makecell{\bfseries plskz362\\N = 362\\M = 880} &
\imgcell{figures/large_graphs/28_sgd2.pdf} &
\imgcell{figures/large_graphs/28_pmds.pdf} &
\imgcell{figures/large_graphs/28_fa2.pdf} &
\imgcell{figures/large_graphs/28_deepgd.pdf} &
\imgcell{figures/large_graphs/28_gd2_stress_xing.pdf} &
\imgcell{figures/large_graphs/28_smartgd_stress.pdf} &
\imgcell{figures/large_graphs/28_smartgd_xing.pdf} &
\imgcell{figures/large_graphs/28_smartgd_xing_nsc.pdf} &
\imgcell{figures/large_graphs/28_smartgd_xangle.pdf} &
\imgcell{figures/large_graphs/28_smartgd_stress_xing.pdf} &
\imgcell{figures/large_graphs/28_smartgd_stress_xangle.pdf} &
\imgcell{figures/large_graphs/28_smartgd_combined.pdf} \\

&
t = 0.04s &
t = 4.47s &
t = 10.02s &
t = 0.64s &
t = 7200.00s &
t = 0.60s &
t = 0.49s &
t = 0.57s &
t = 0.58s &
t = 0.42s &
t = 0.65s &
t = 0.51s \\

\makecell{\bfseries plat362\\N = 362\\M = 2712} &
\imgcell{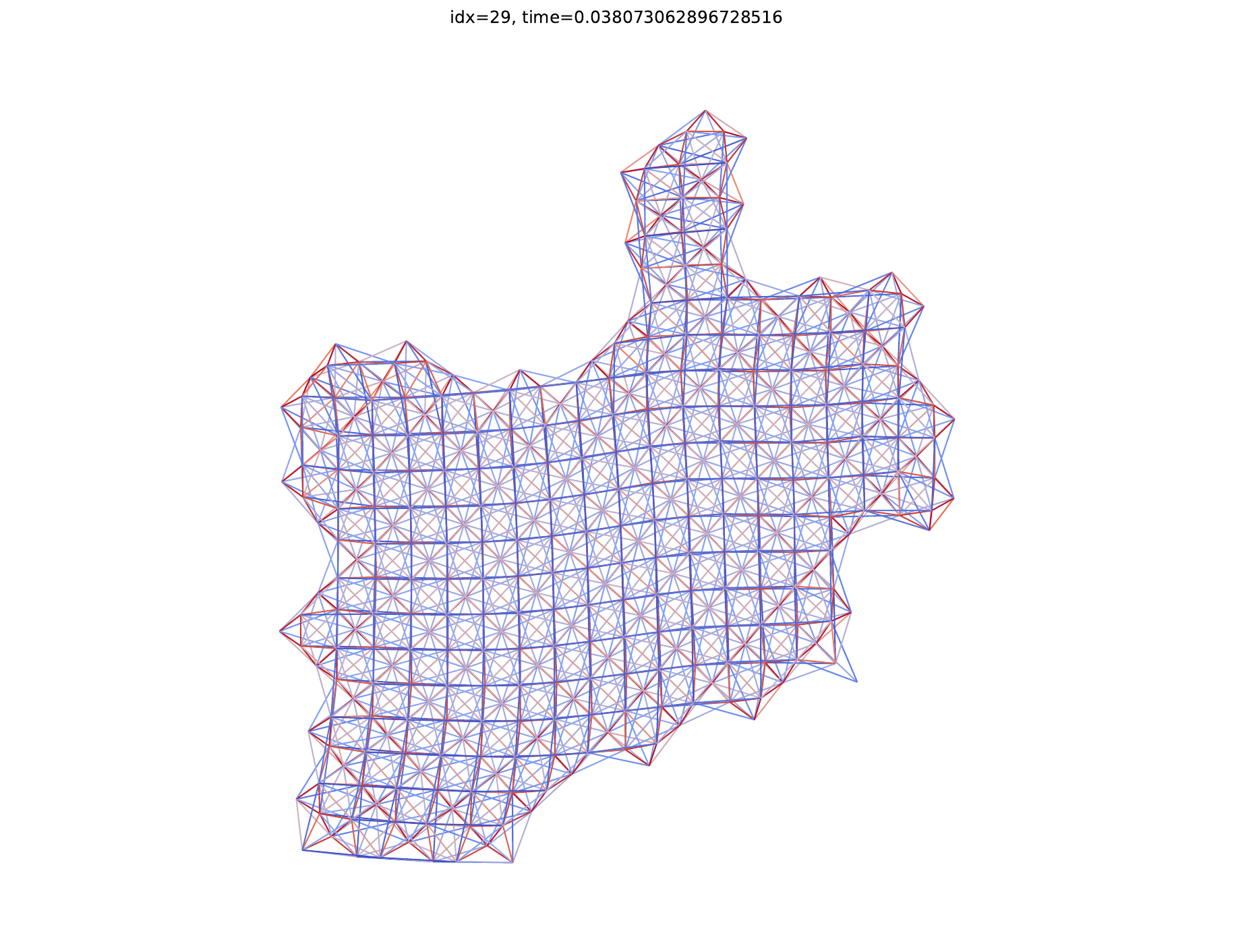} &
\imgcell{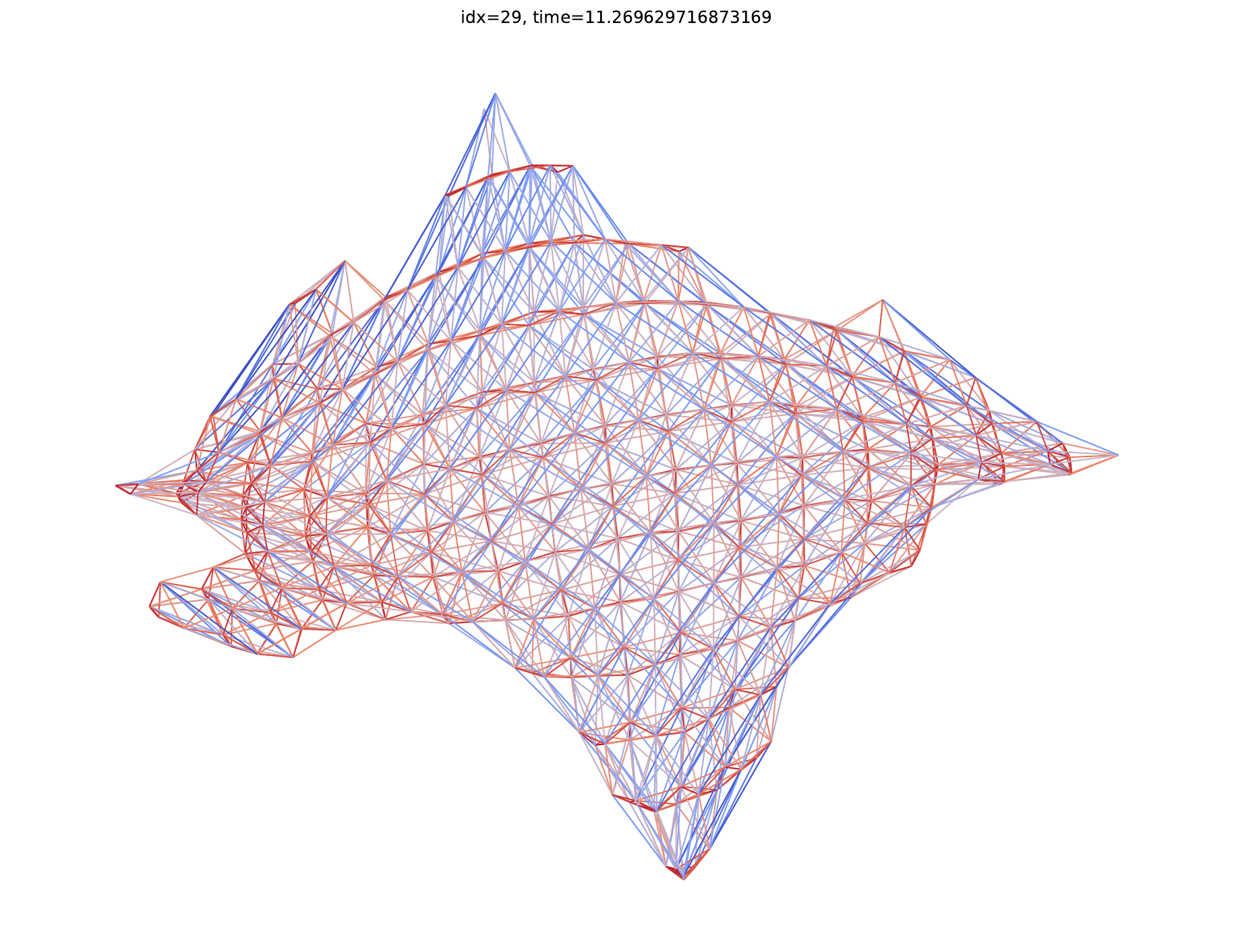} &
\imgcell{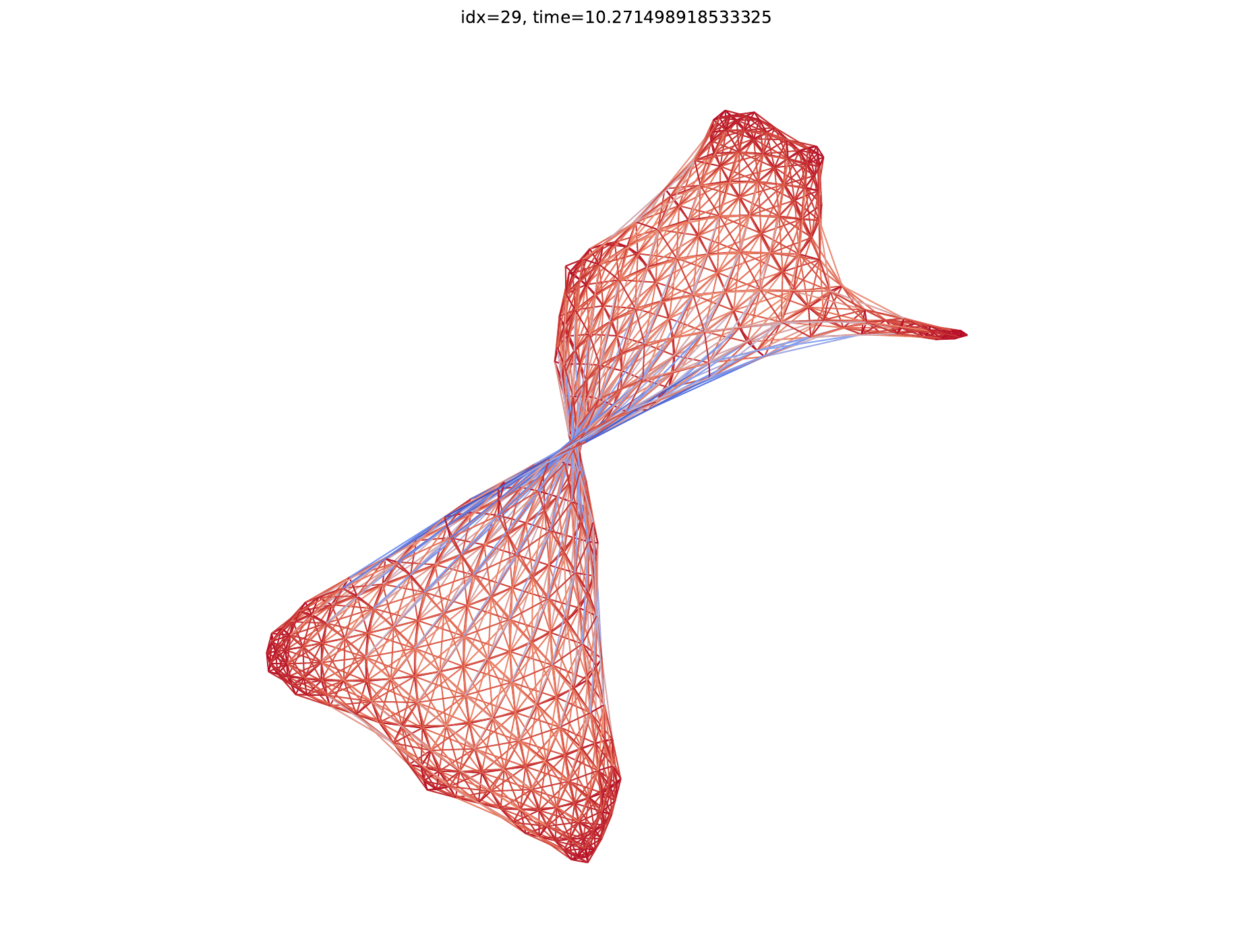} &
\imgcell{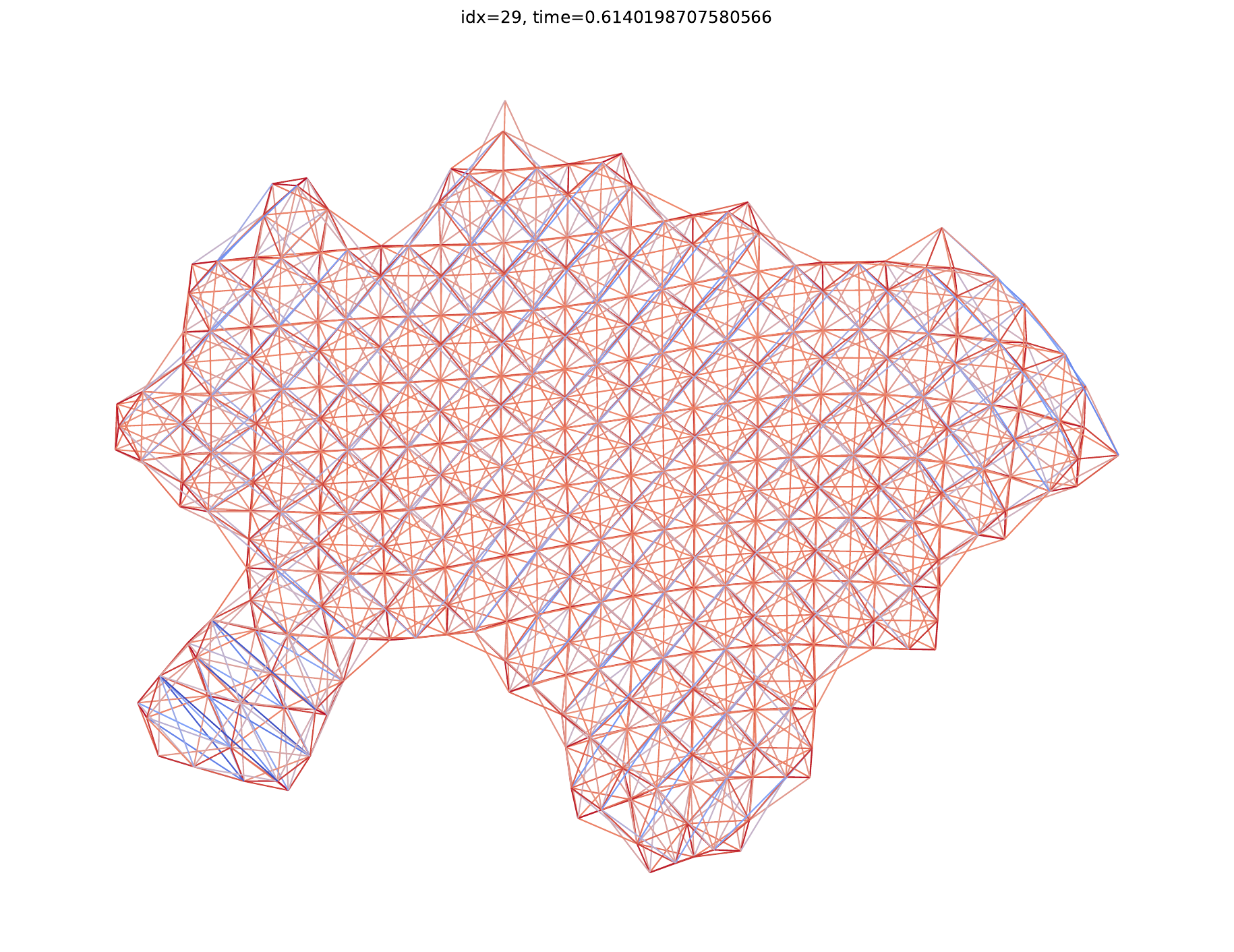} &
\imgcell{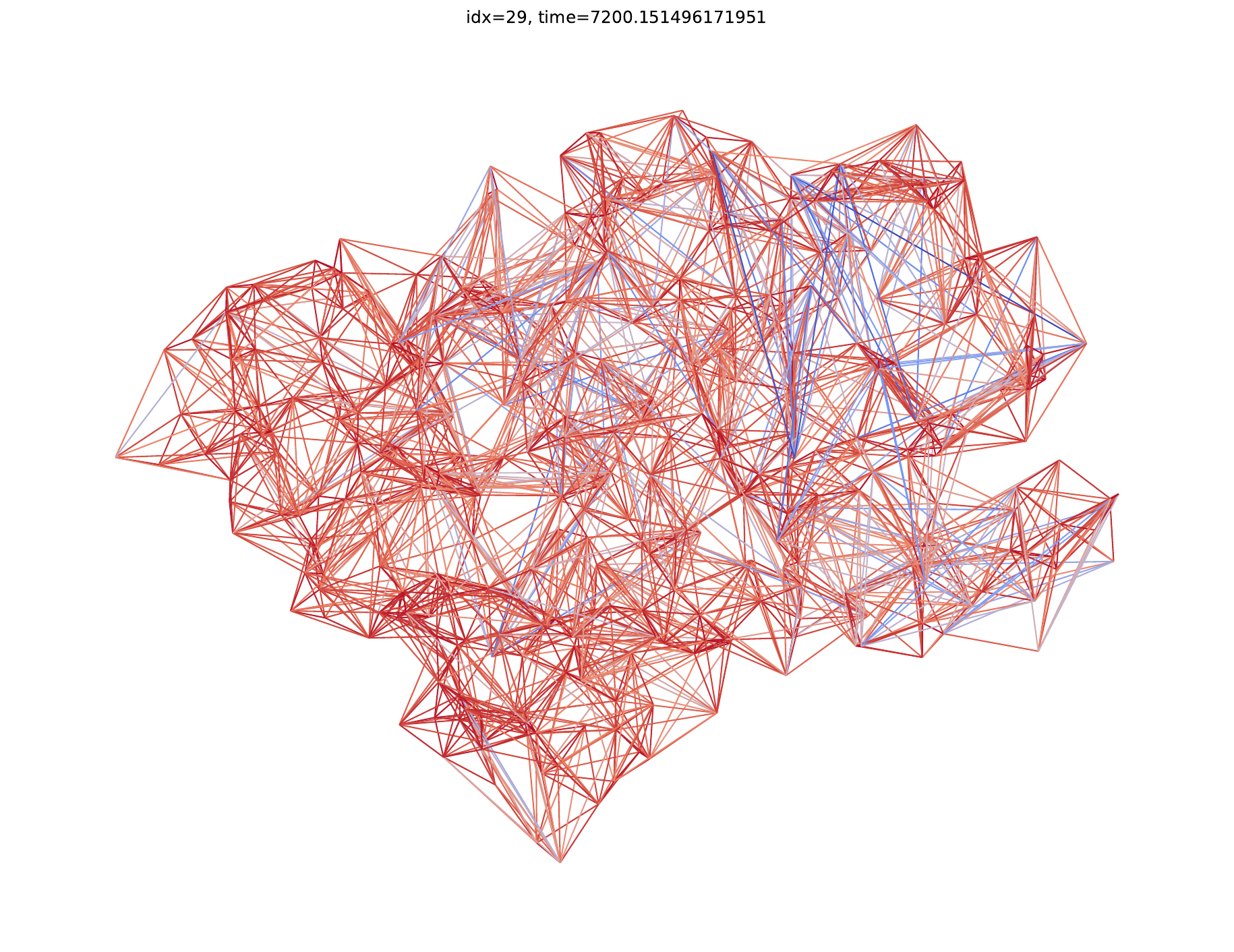} &
\imgcell{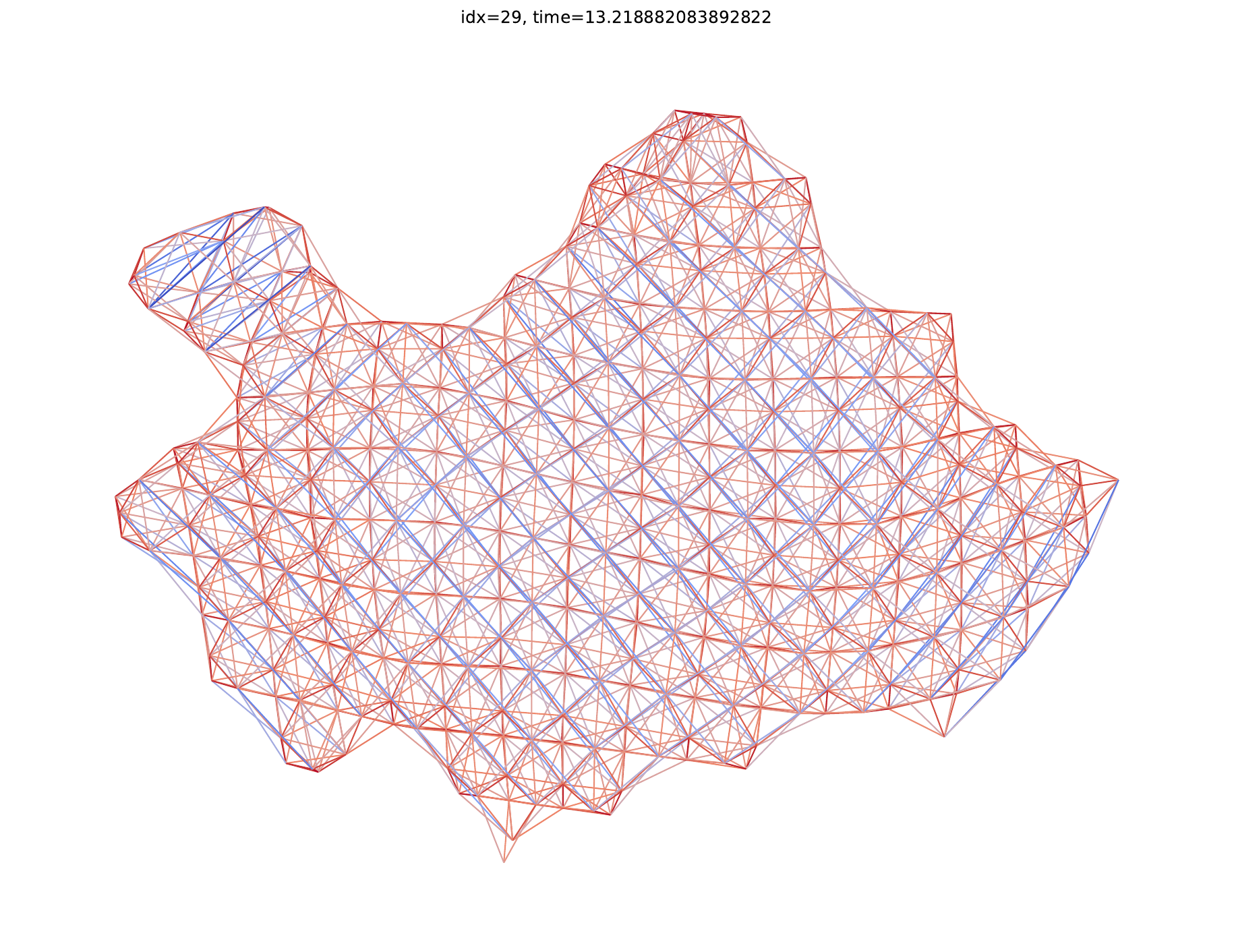} &
\imgcell{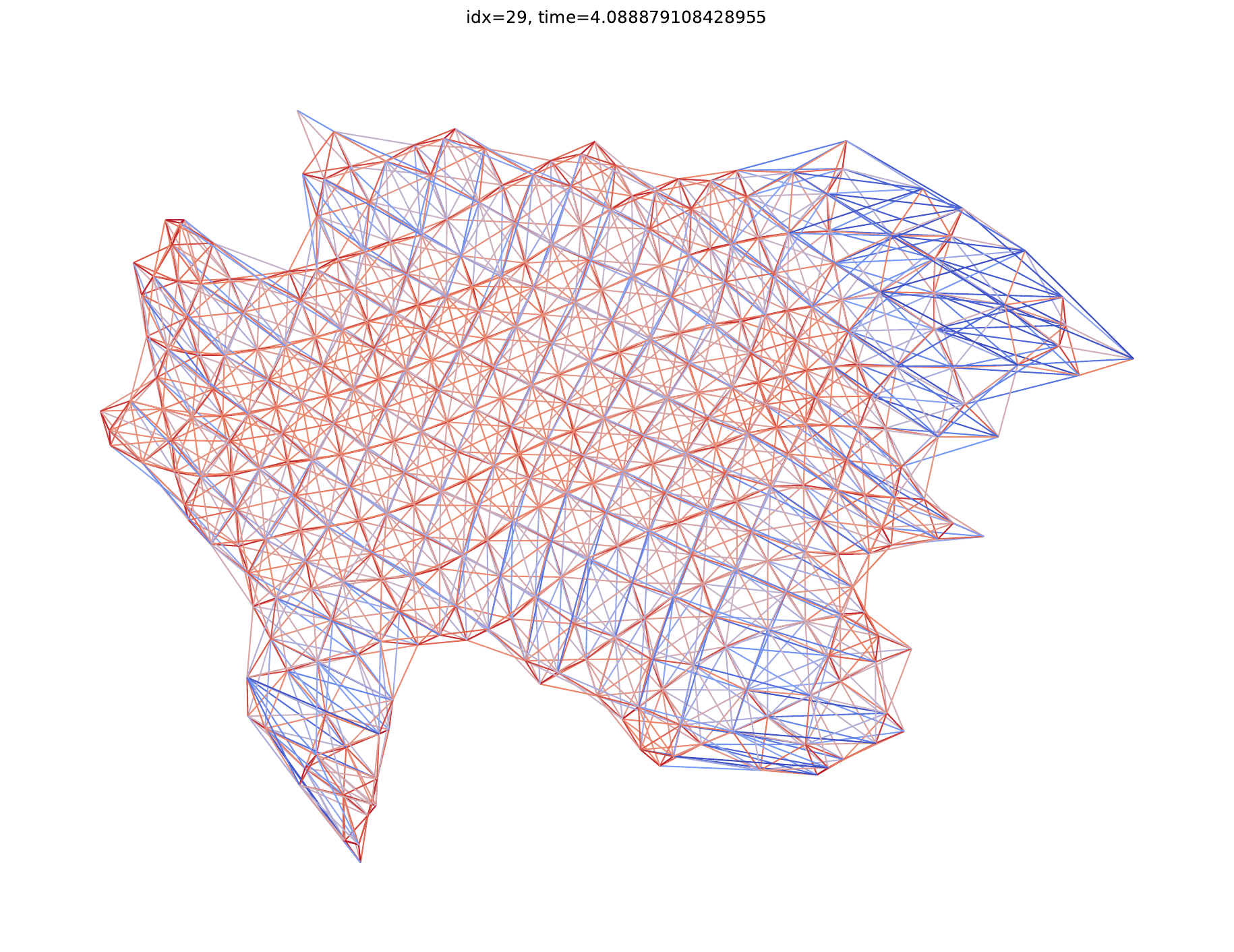} &
\imgcell{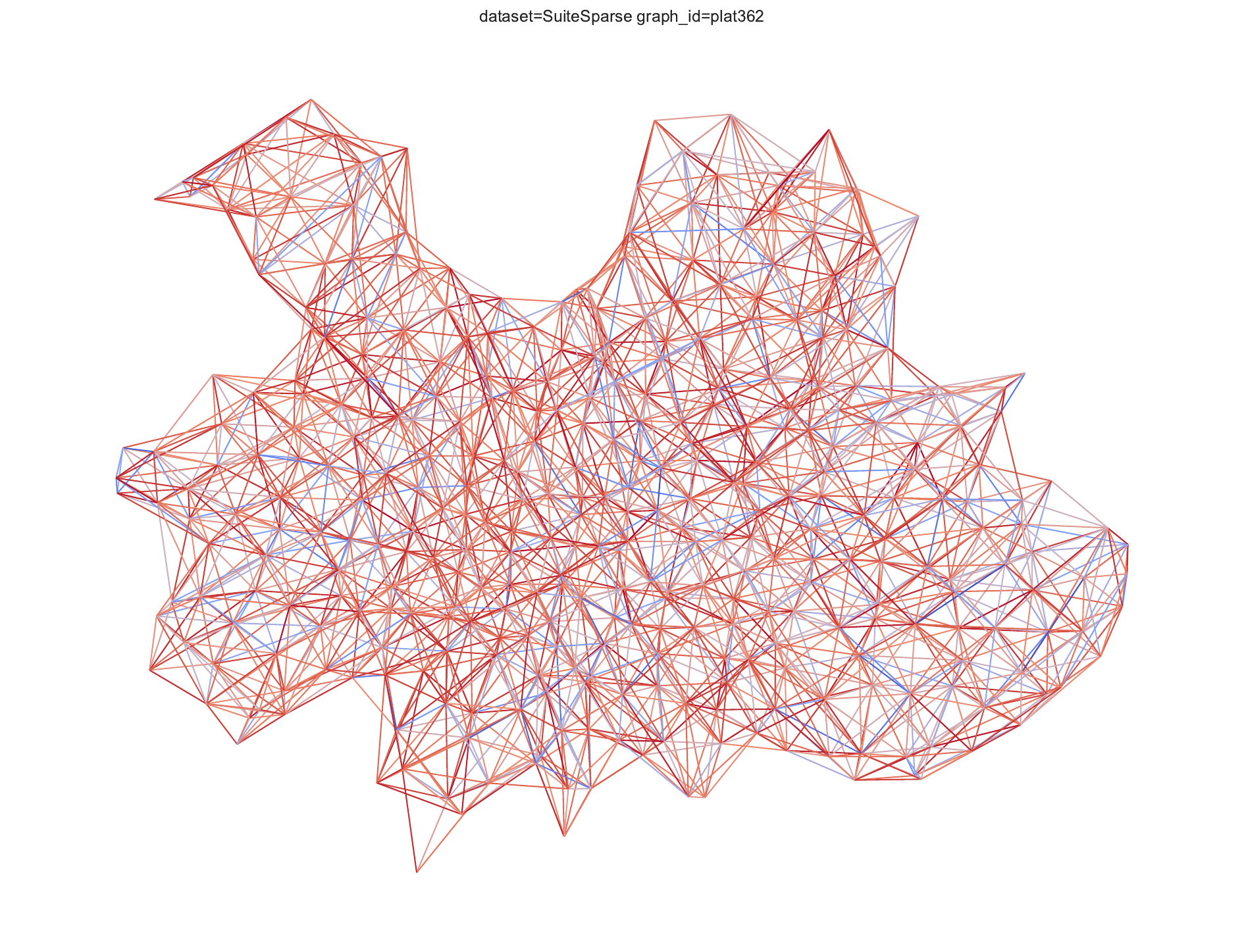} &
\imgcell{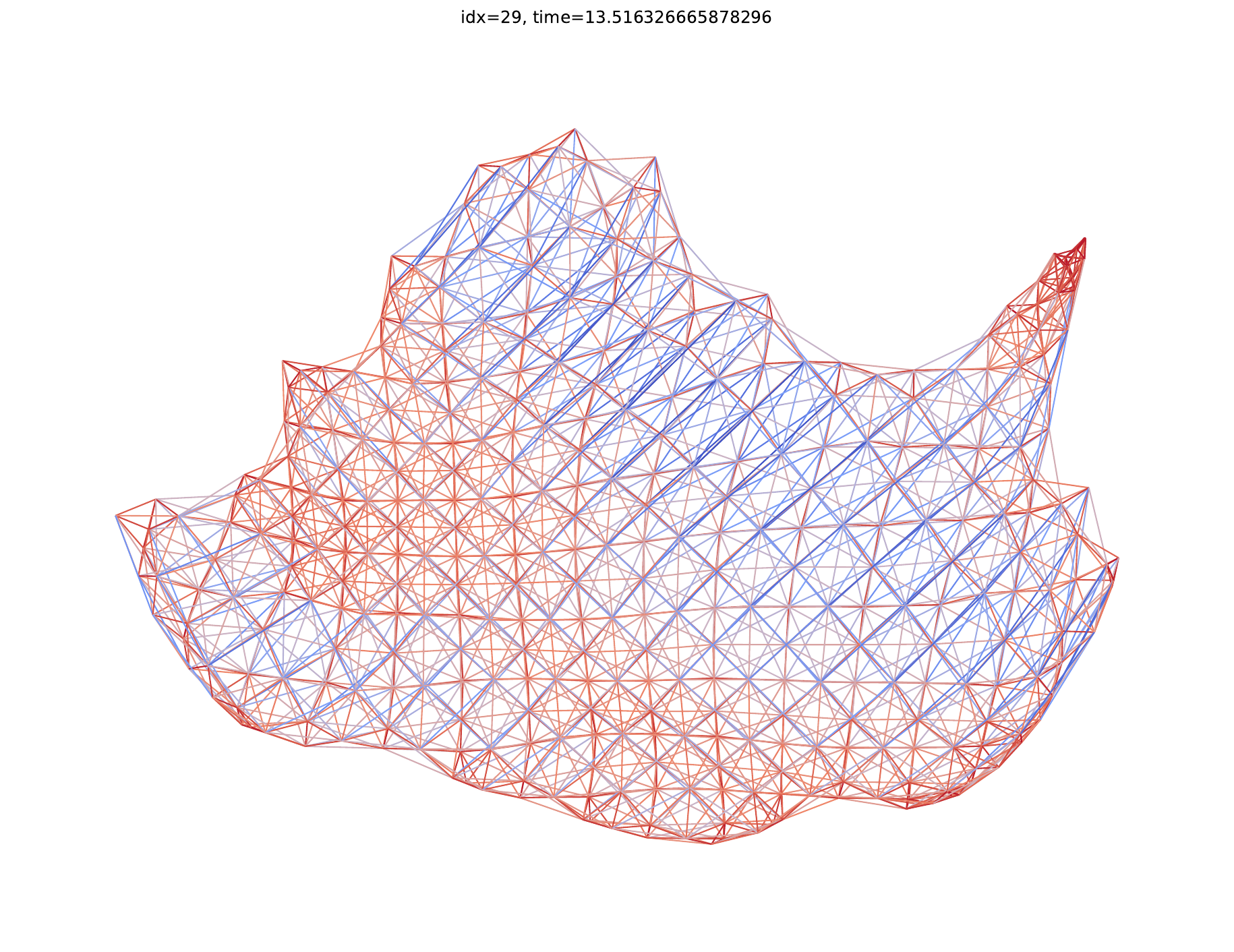} &
\imgcell{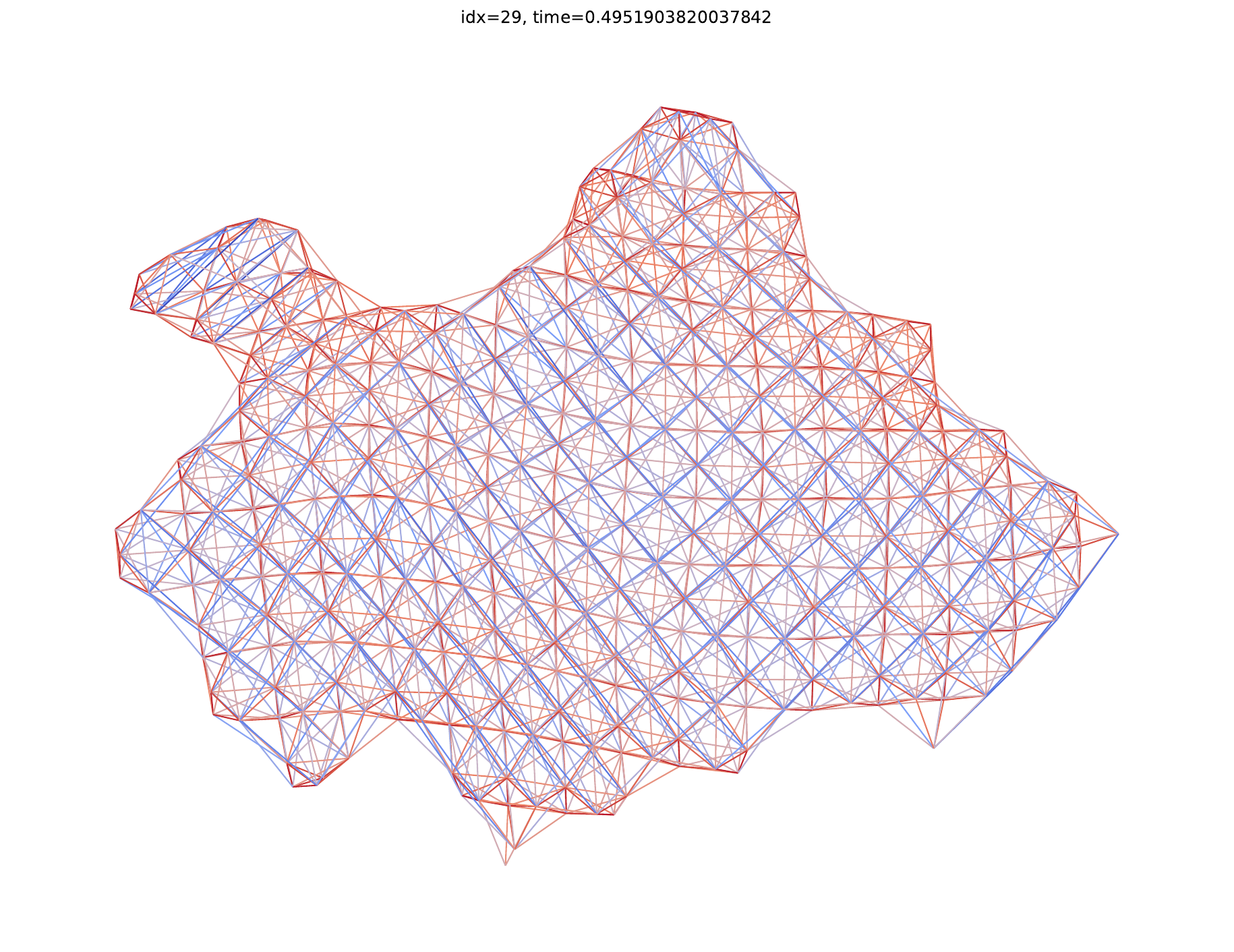} &
\imgcell{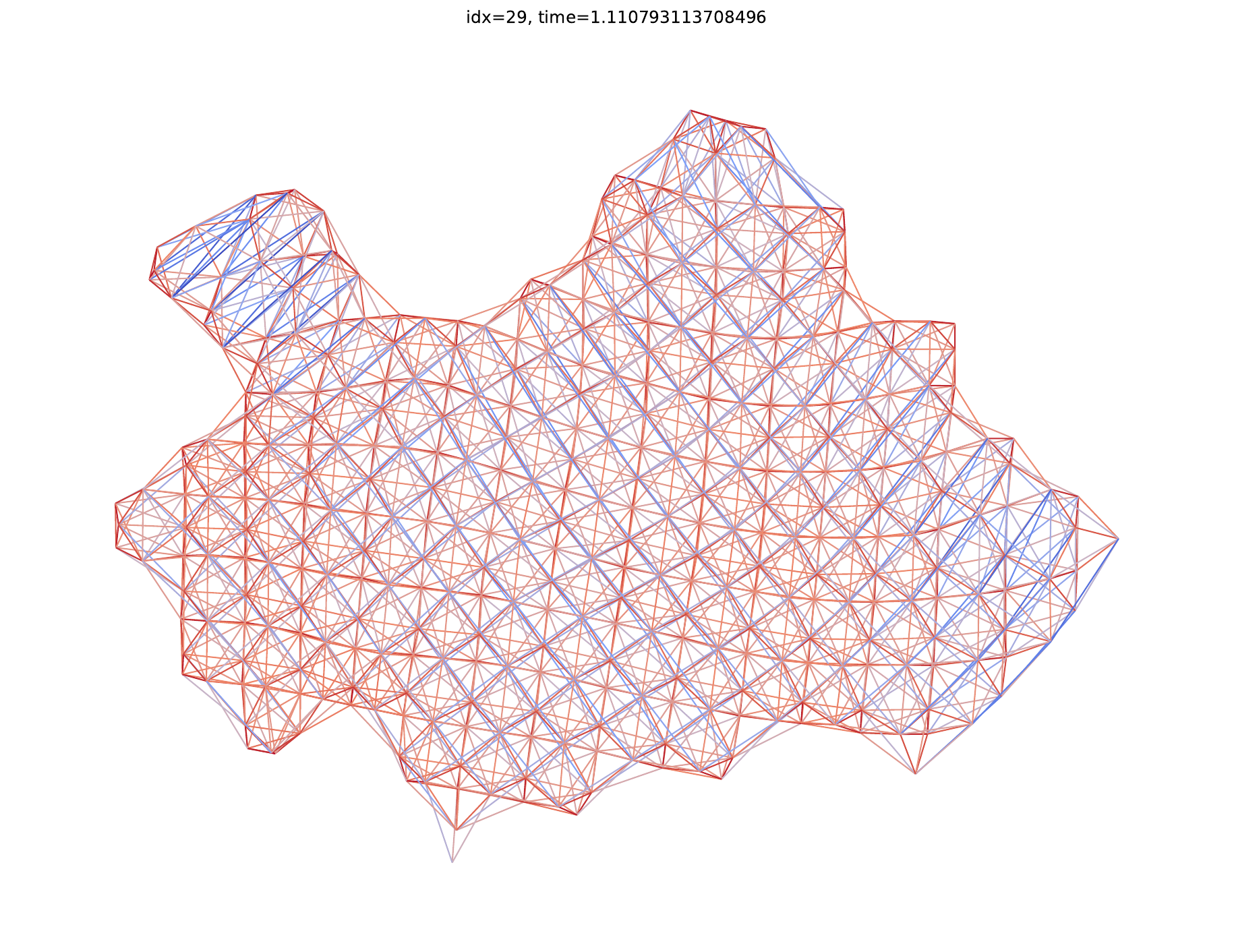} &
\imgcell{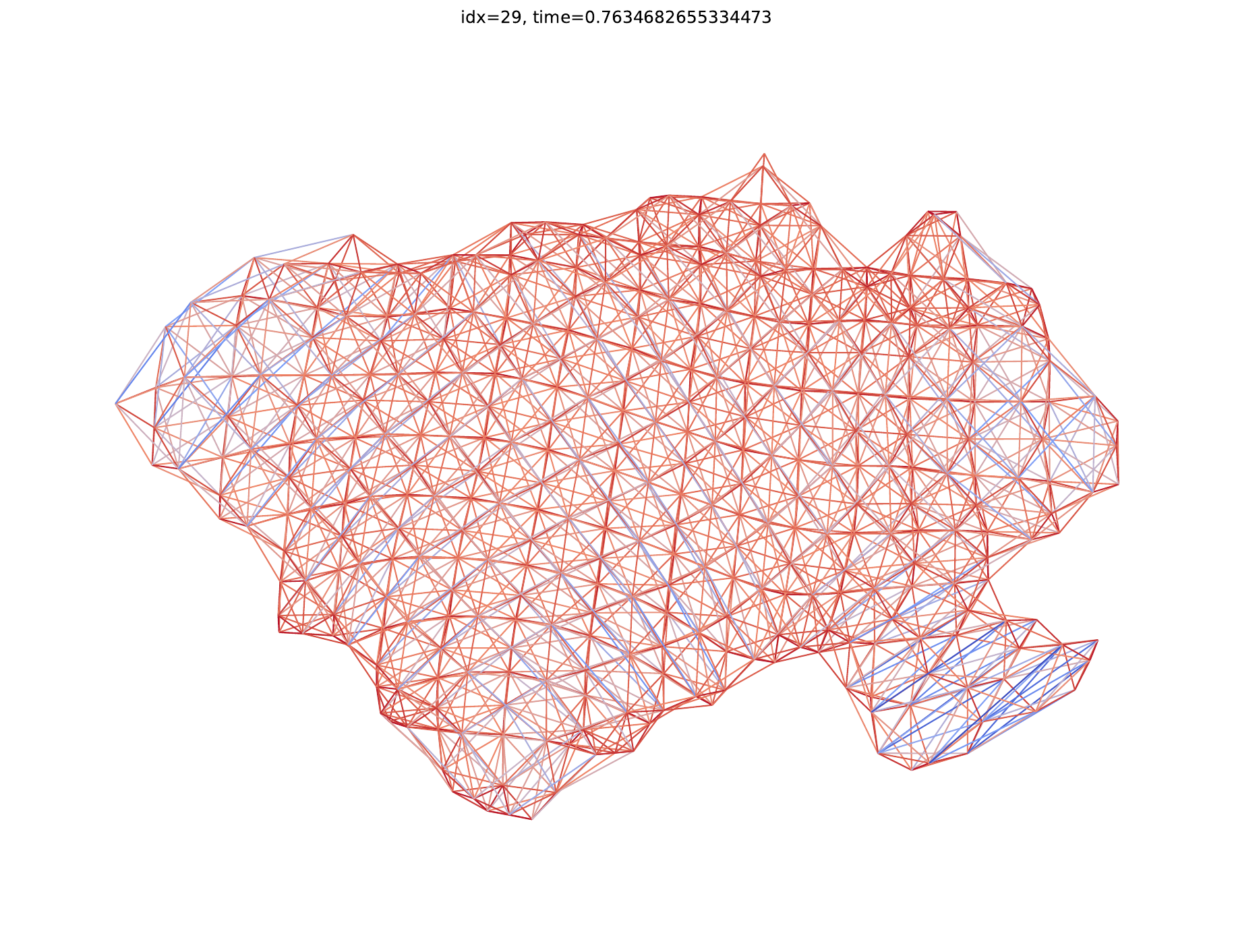} \\

&
t = 0.04s &
t = 11.27s &
t = 10.27s &
t = 0.61s &
t = 7200.00s &
t = 0.77s &
t = 0.75s &
t = 0.58s &
t = 0.62s &
t = 0.56s &
t = 0.59s &
t = 0.71s \\

\makecell{\bfseries poisson2D\\N = 367\\M = 1025} &
\imgcell{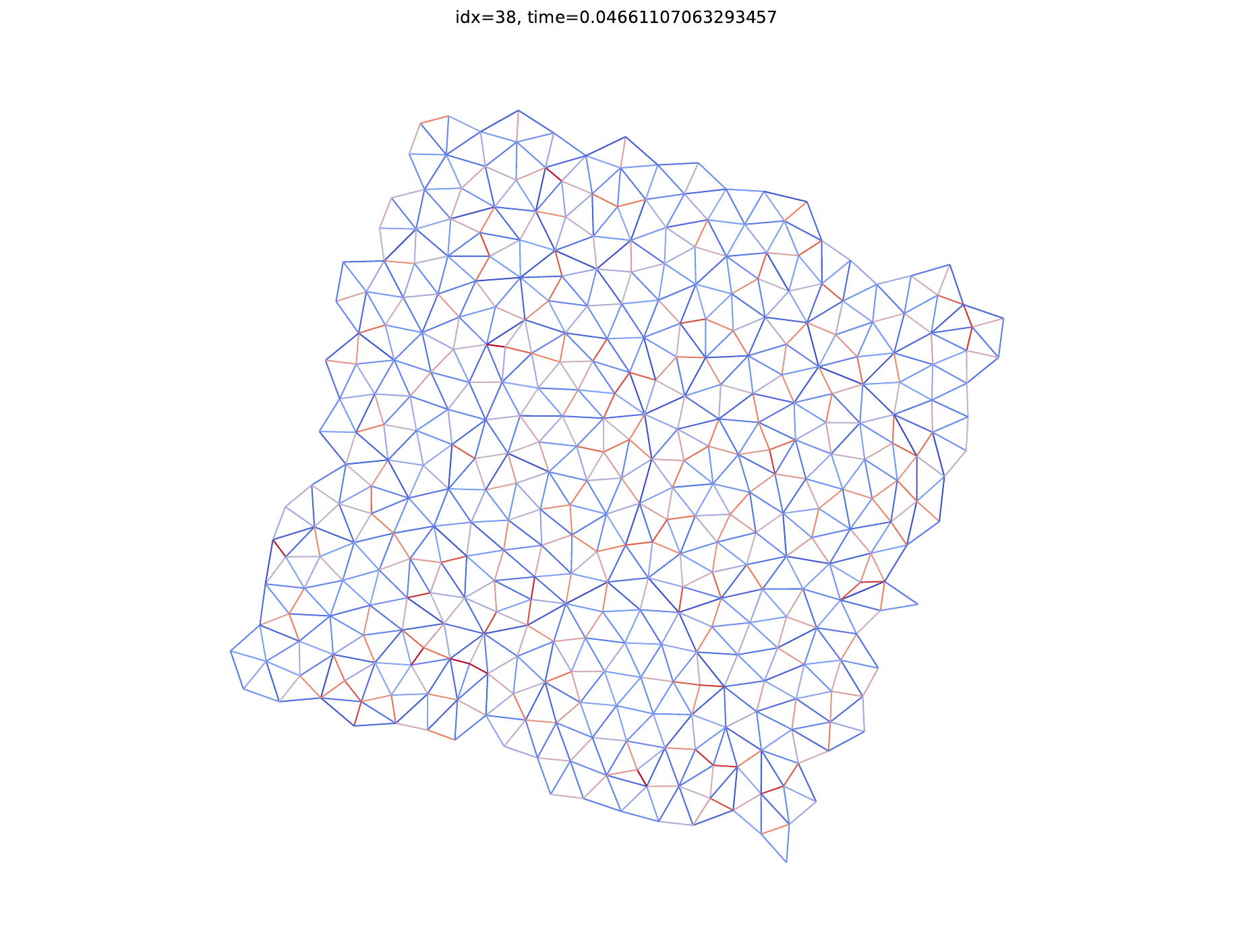} &
\imgcell{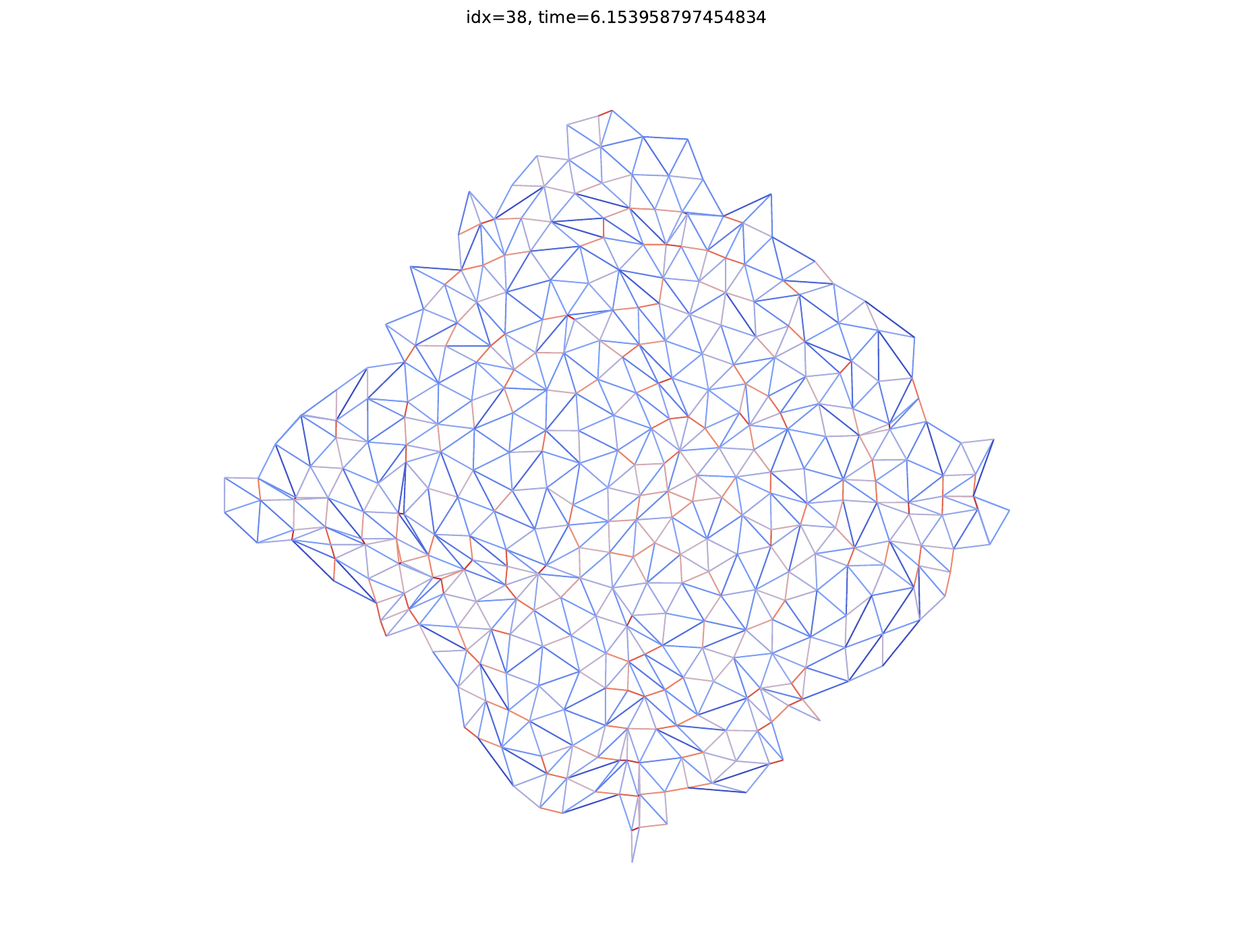} &
\imgcell{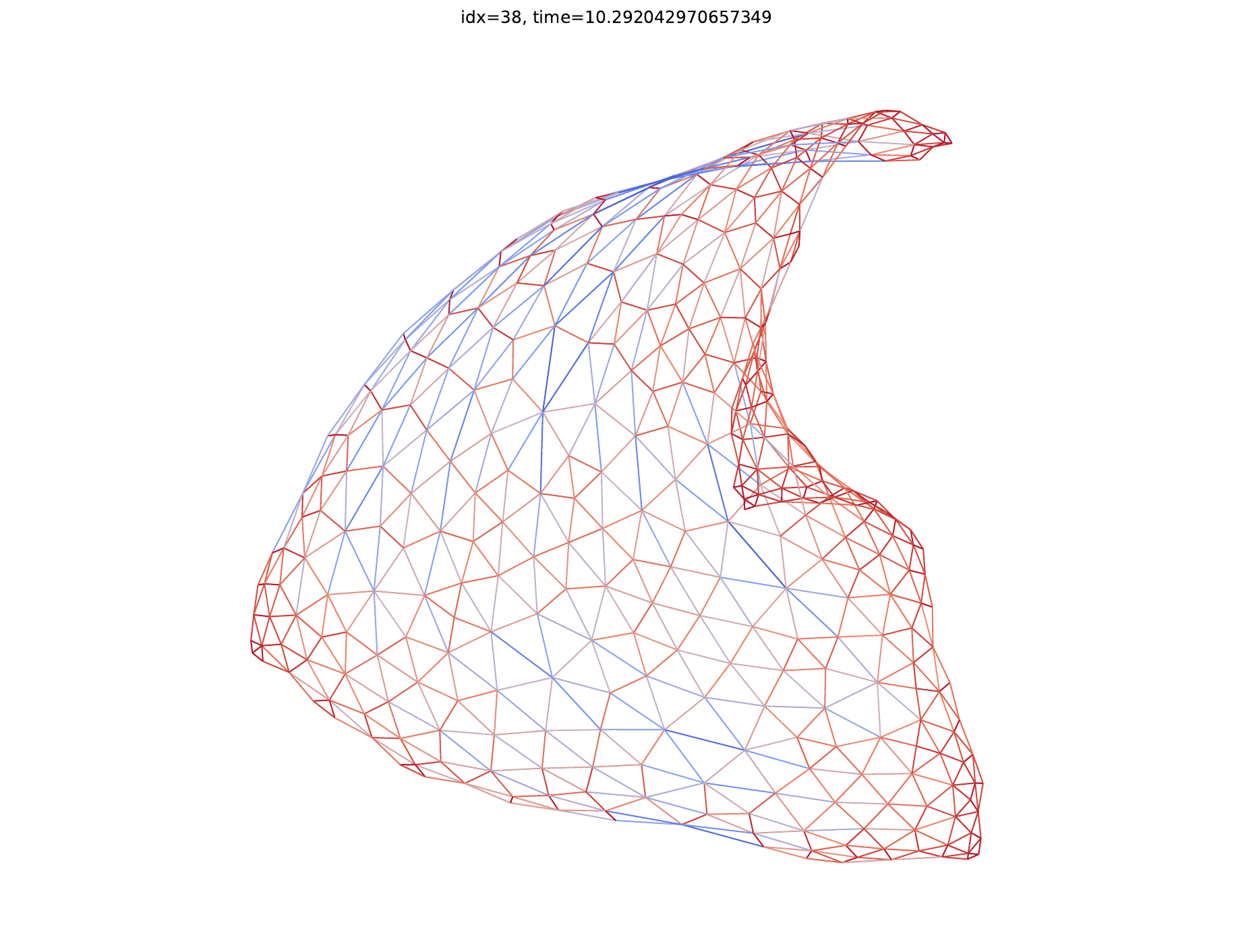} &
\imgcell{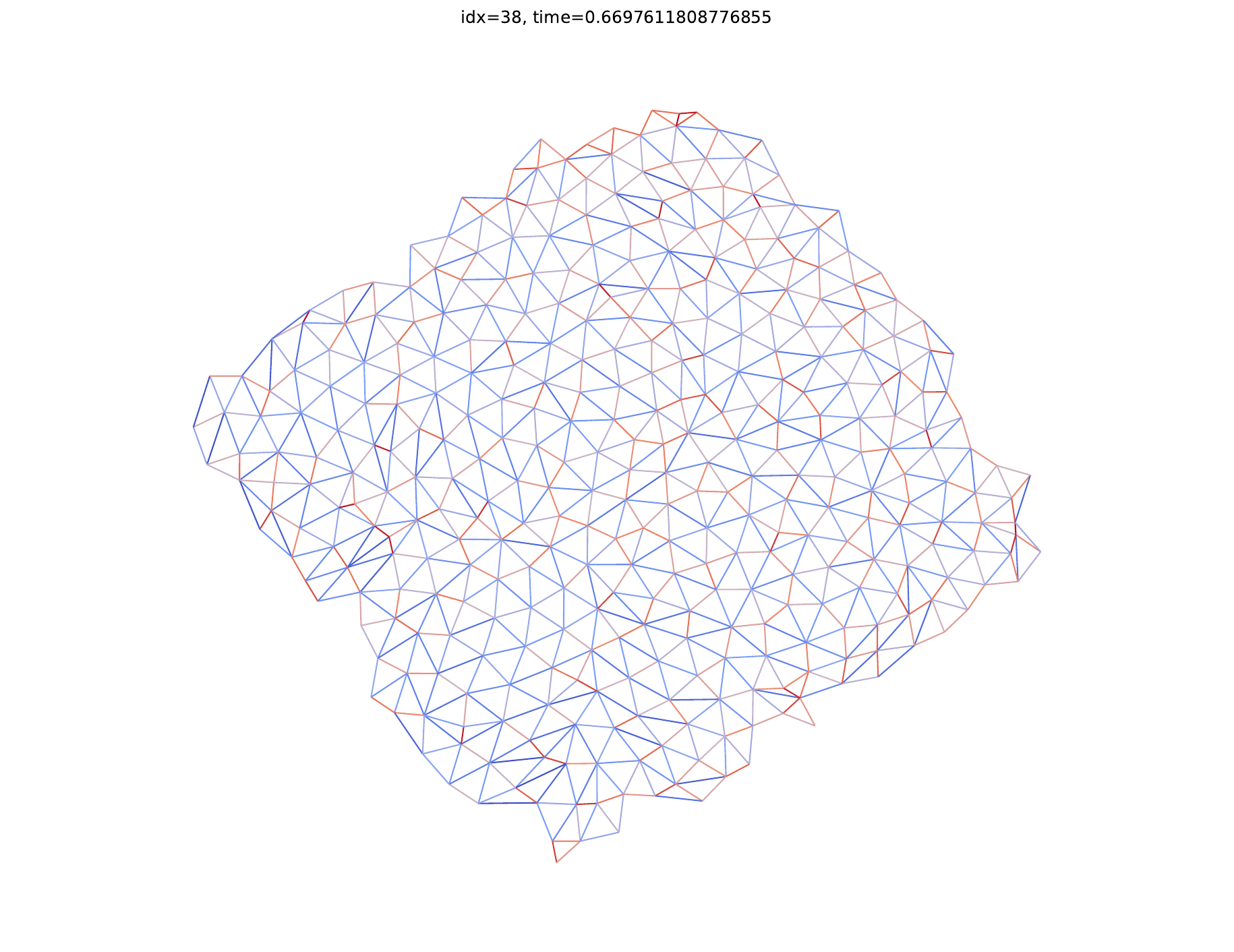} &
\imgcell{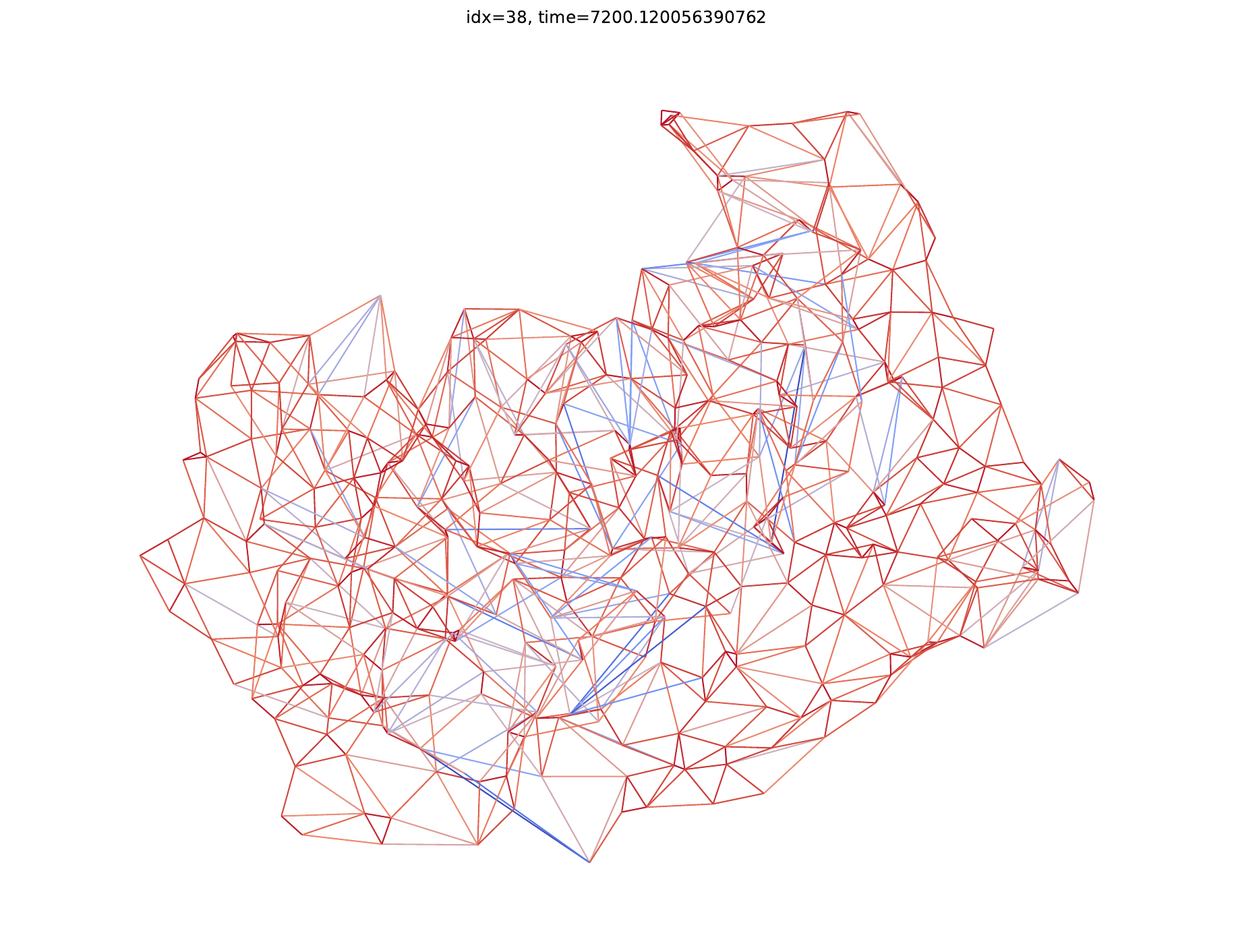} &
\imgcell{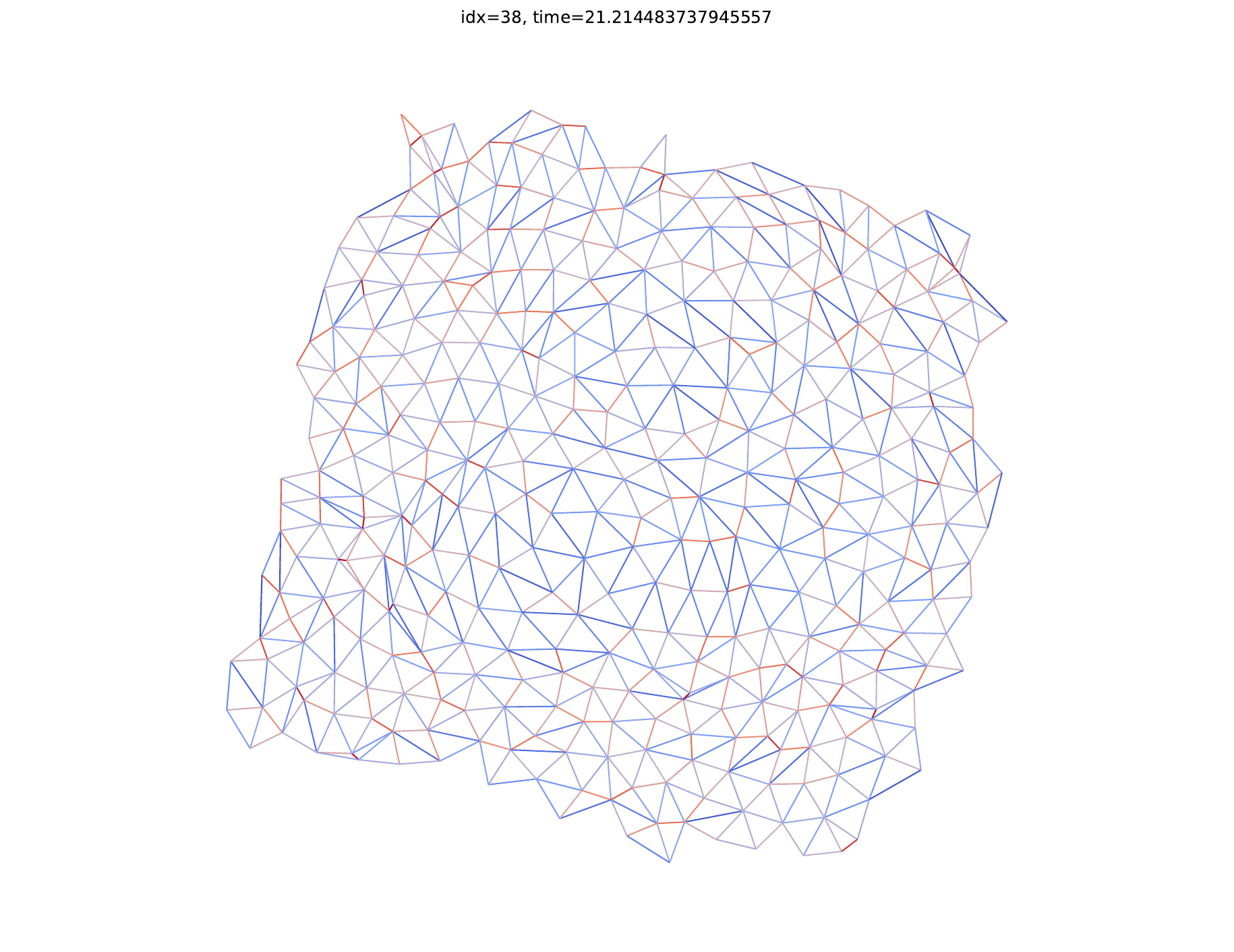} &
\imgcell{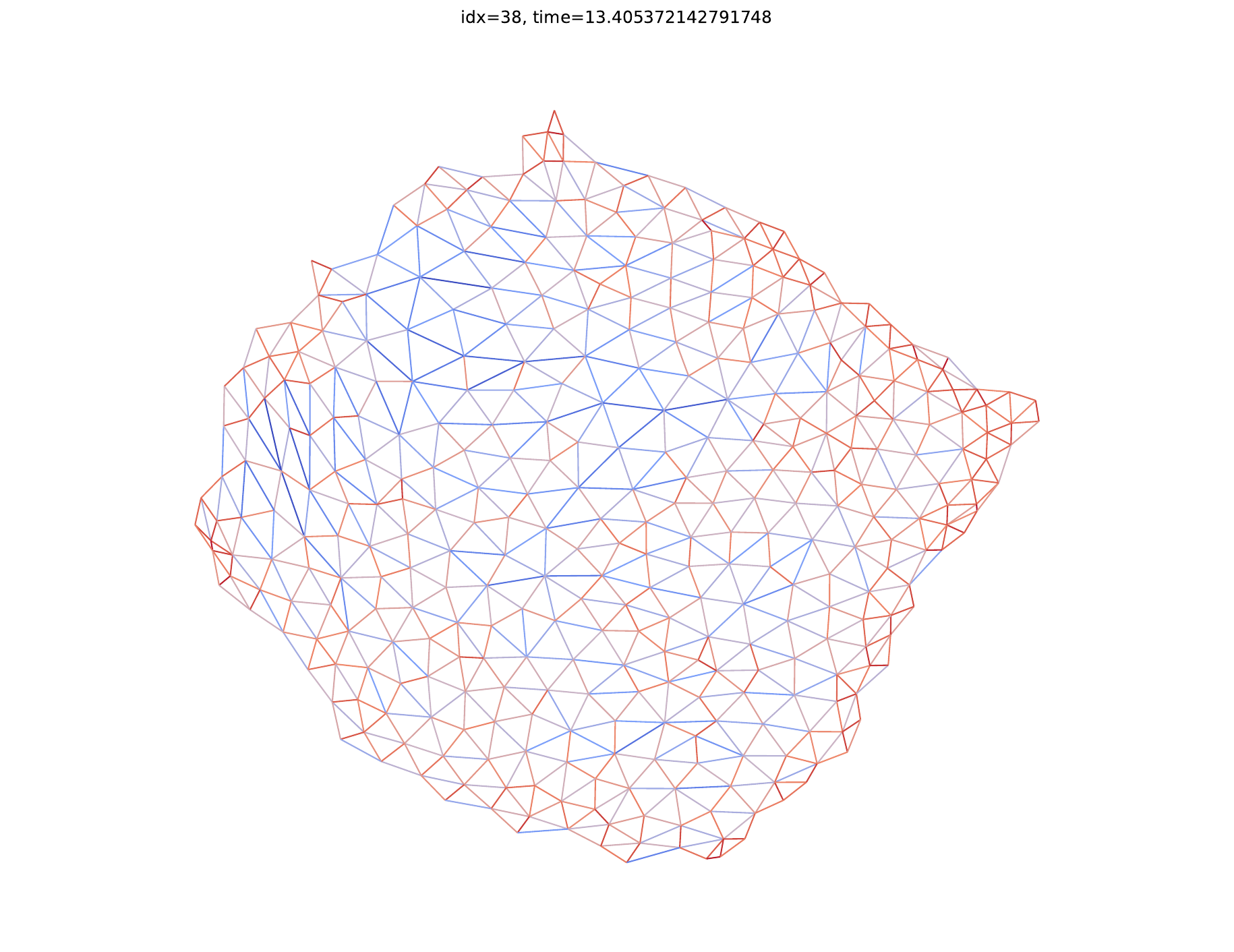} &
\imgcell{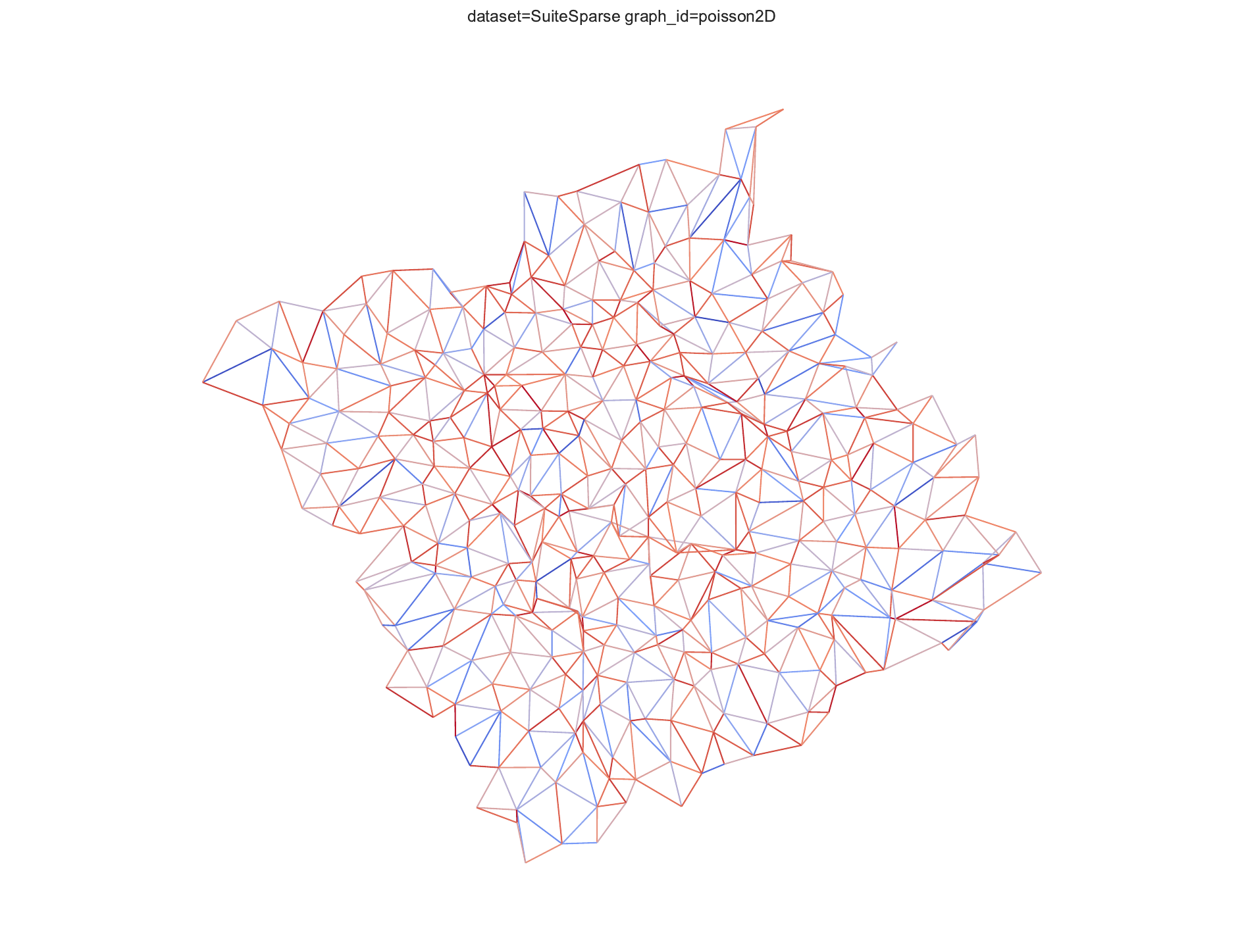} &
\imgcell{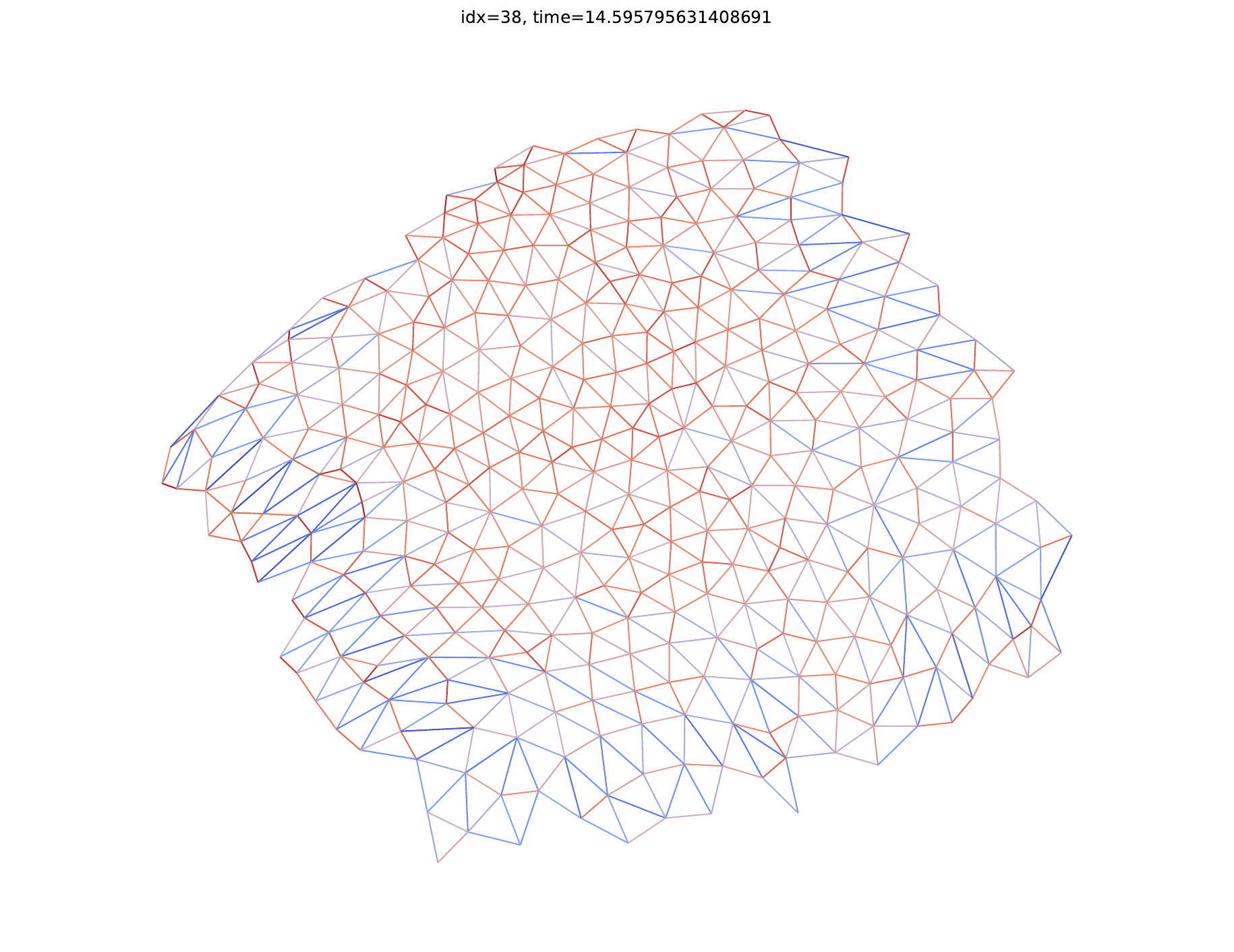} &
\imgcell{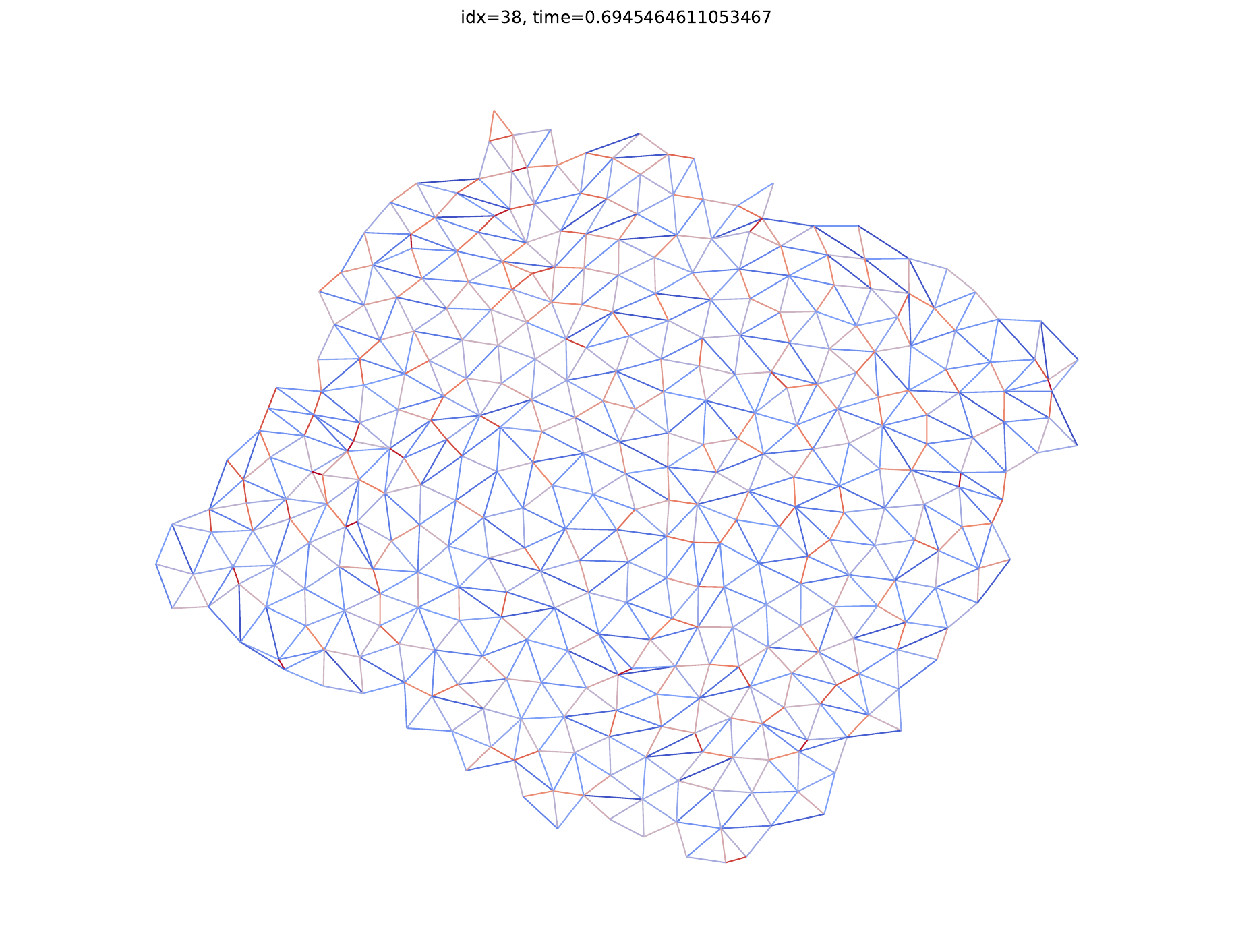} &
\imgcell{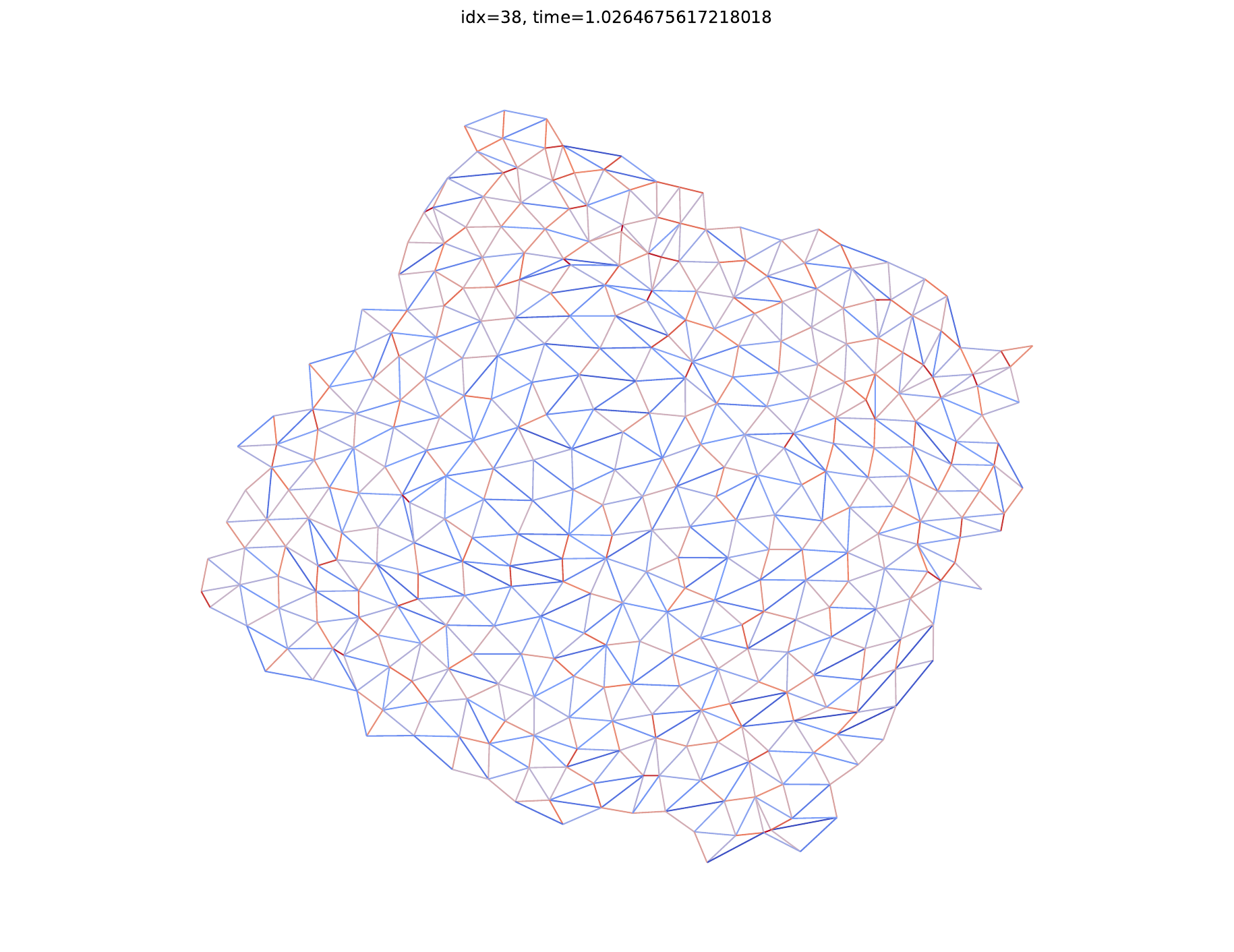} &
\imgcell{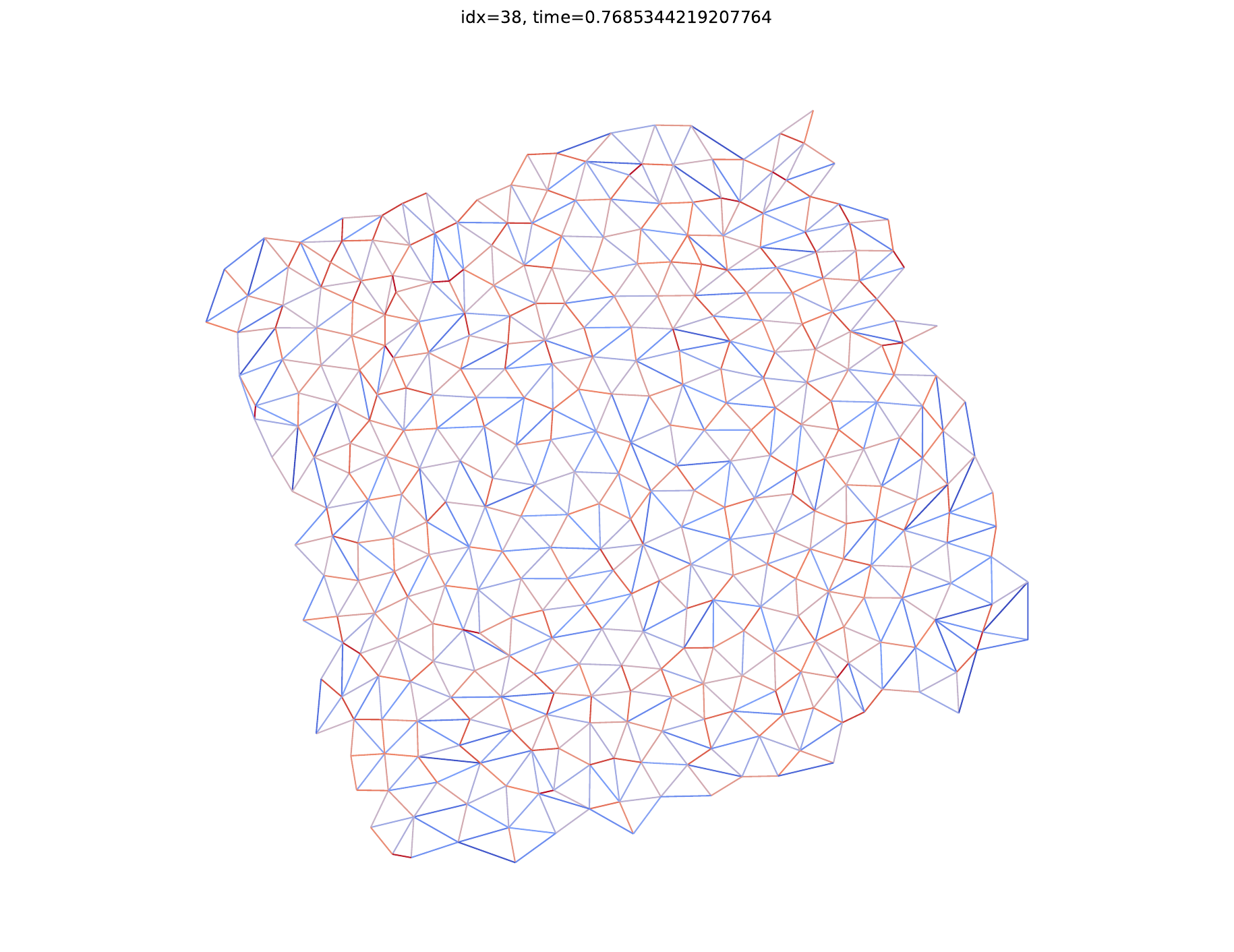} \\

&
t = 0.05s &
t = 6.15s &
t = 10.29s &
t = 0.67s &
t = 7200.00s &
t = 0.79s &
t = 0.72s &
t = 0.66s &
t = 0.69s &
t = 0.53s &
t = 0.63s &
t = 0.50s \\

\makecell{\bfseries bfwa398\\N = 398\\M = 1658} &
\imgcell{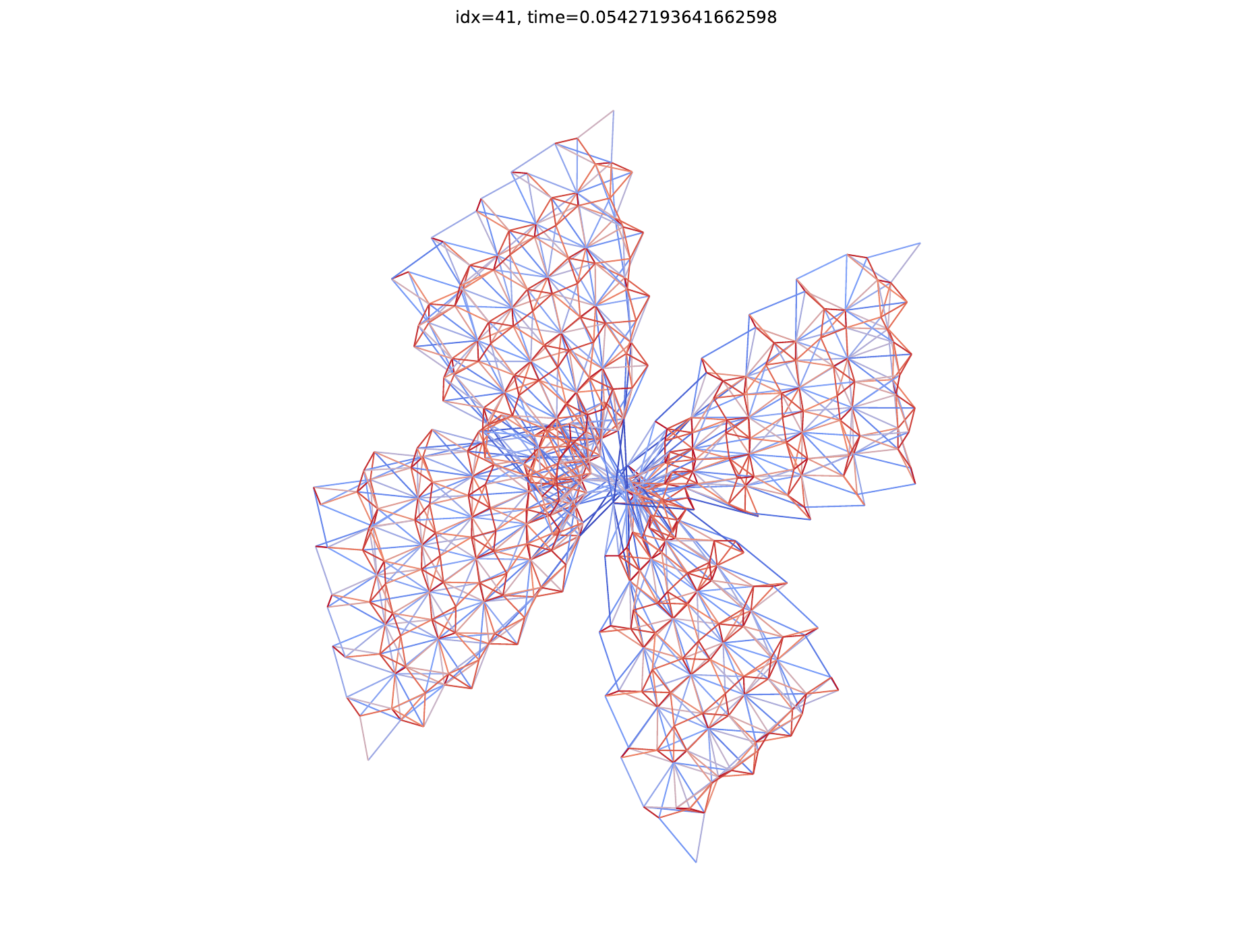} &
\imgcell{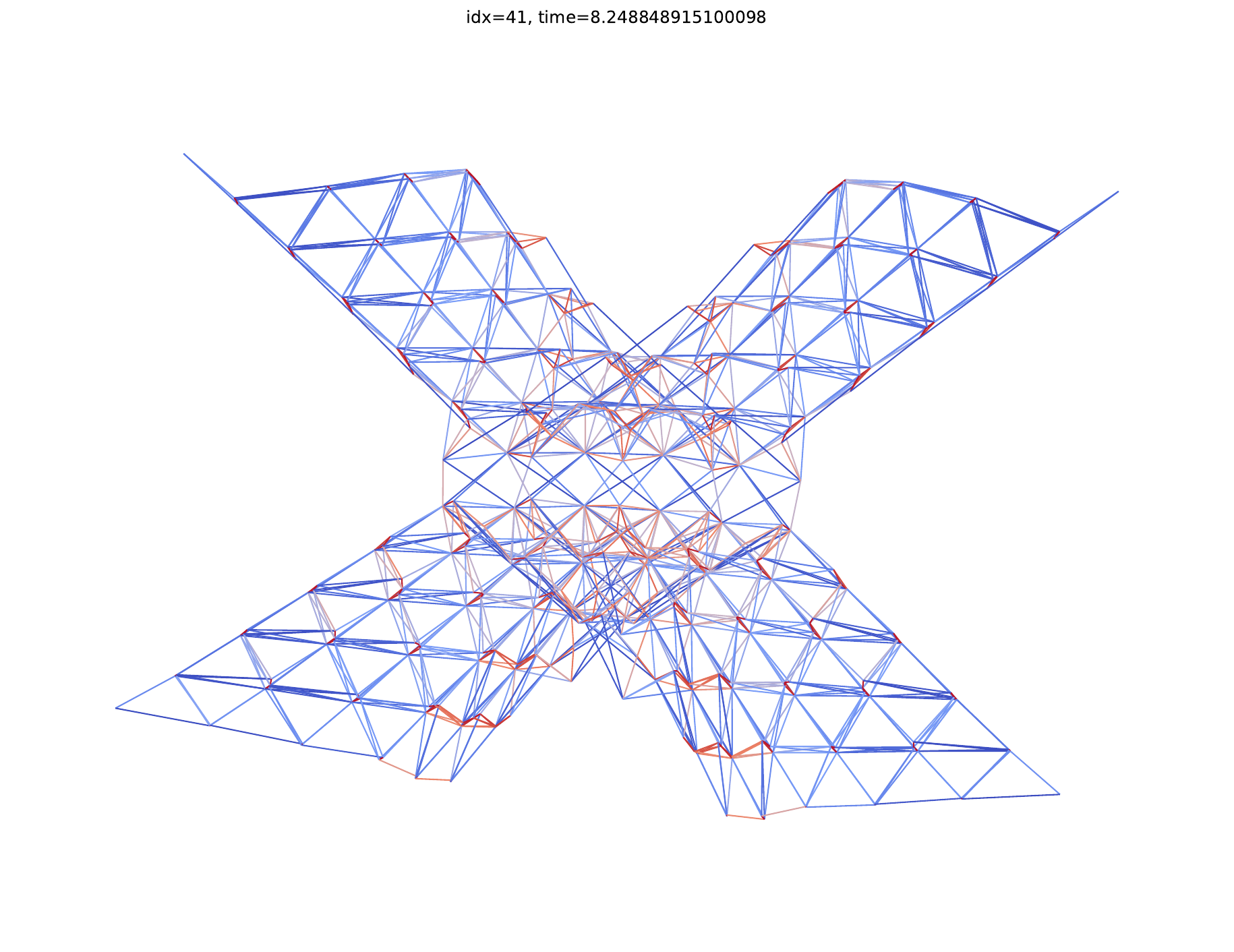} &
\imgcell{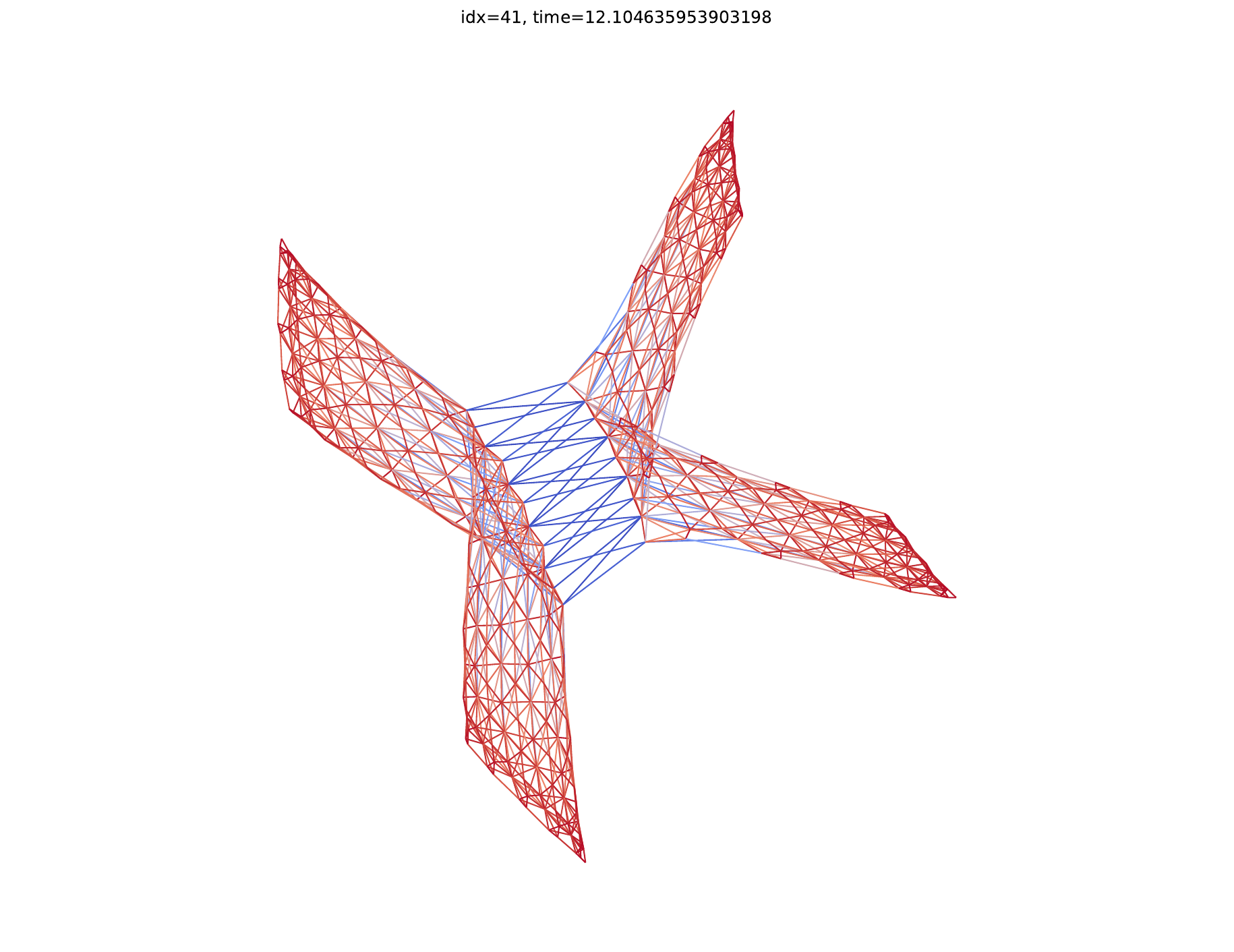} &
\imgcell{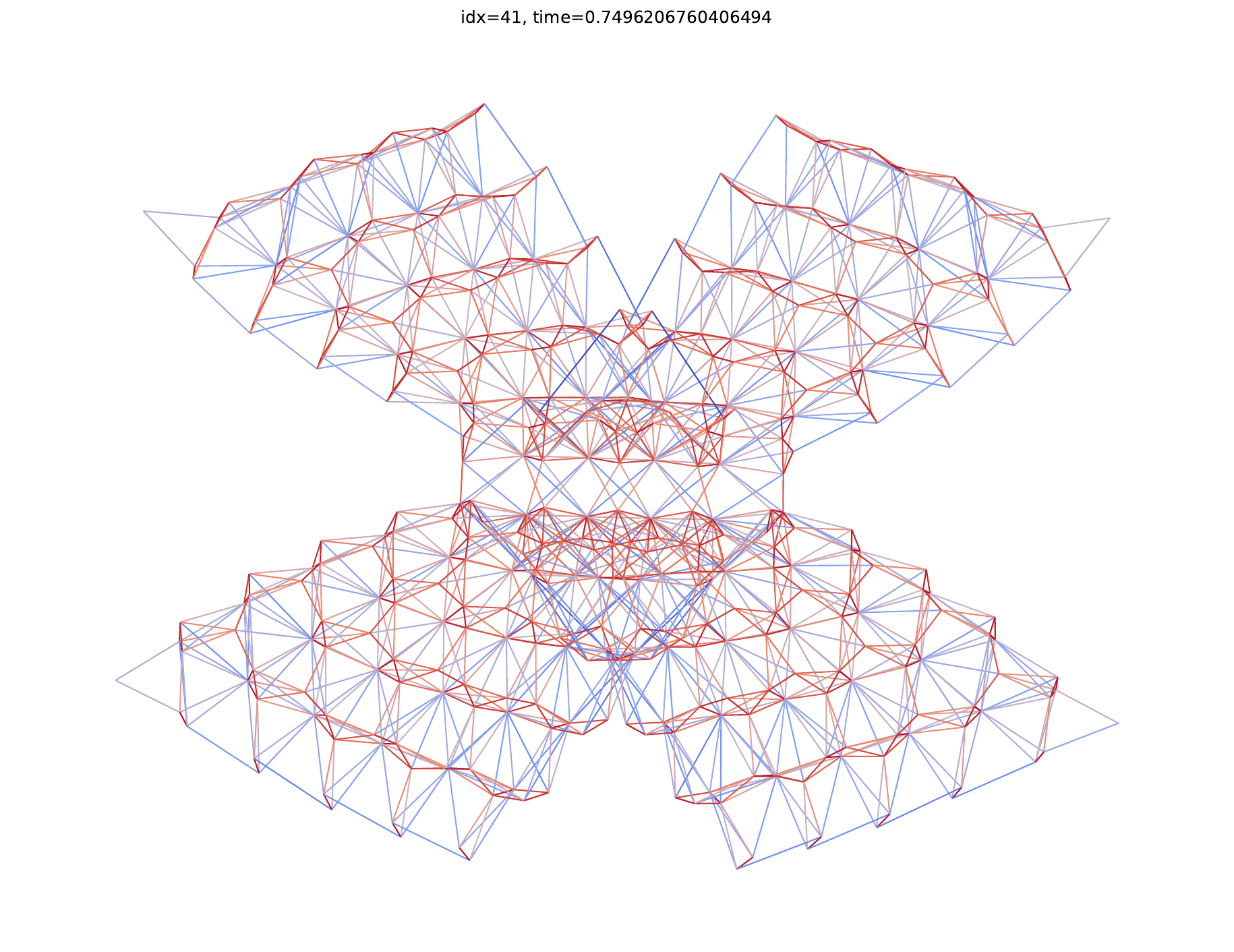} &
\imgcell{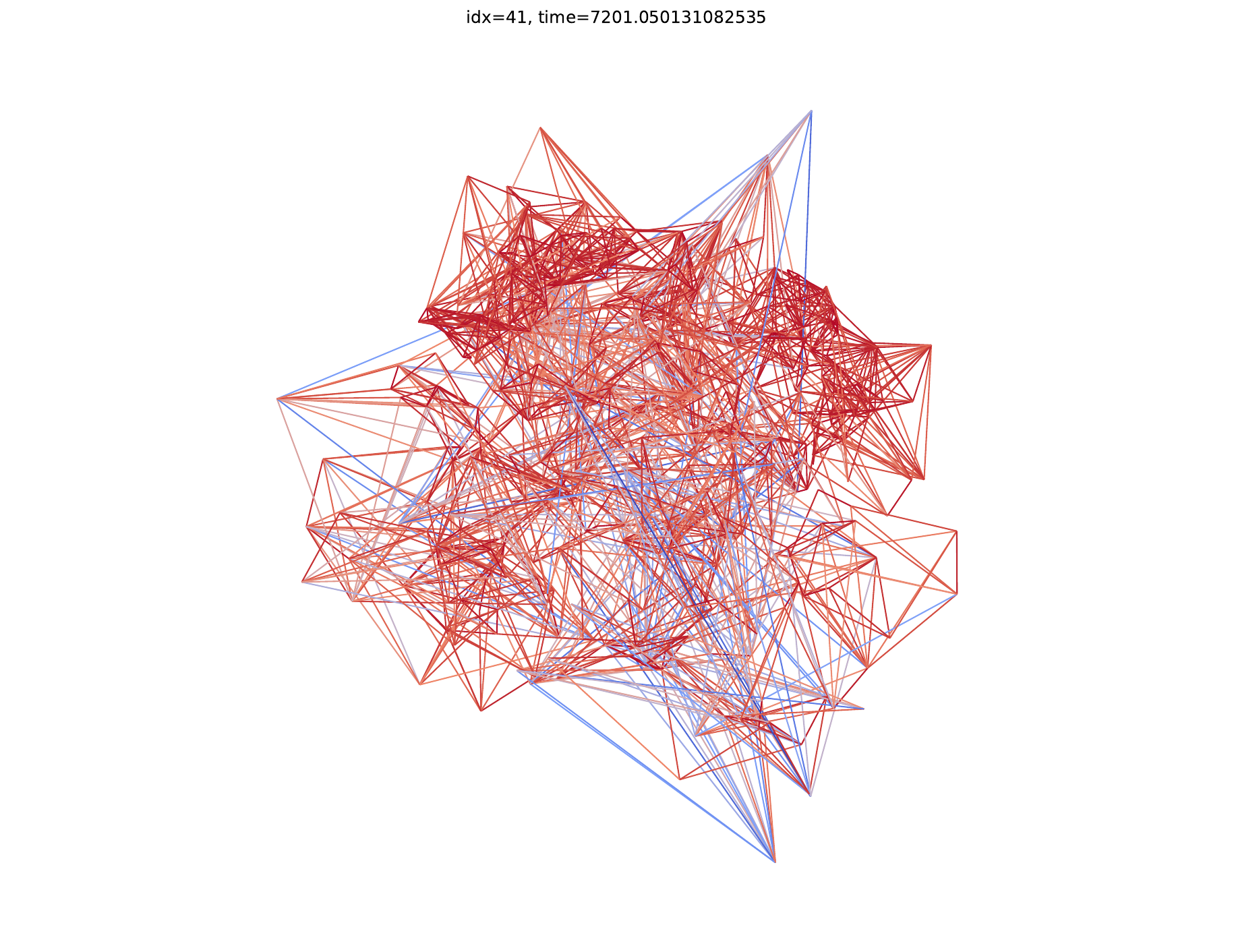} &
\imgcell{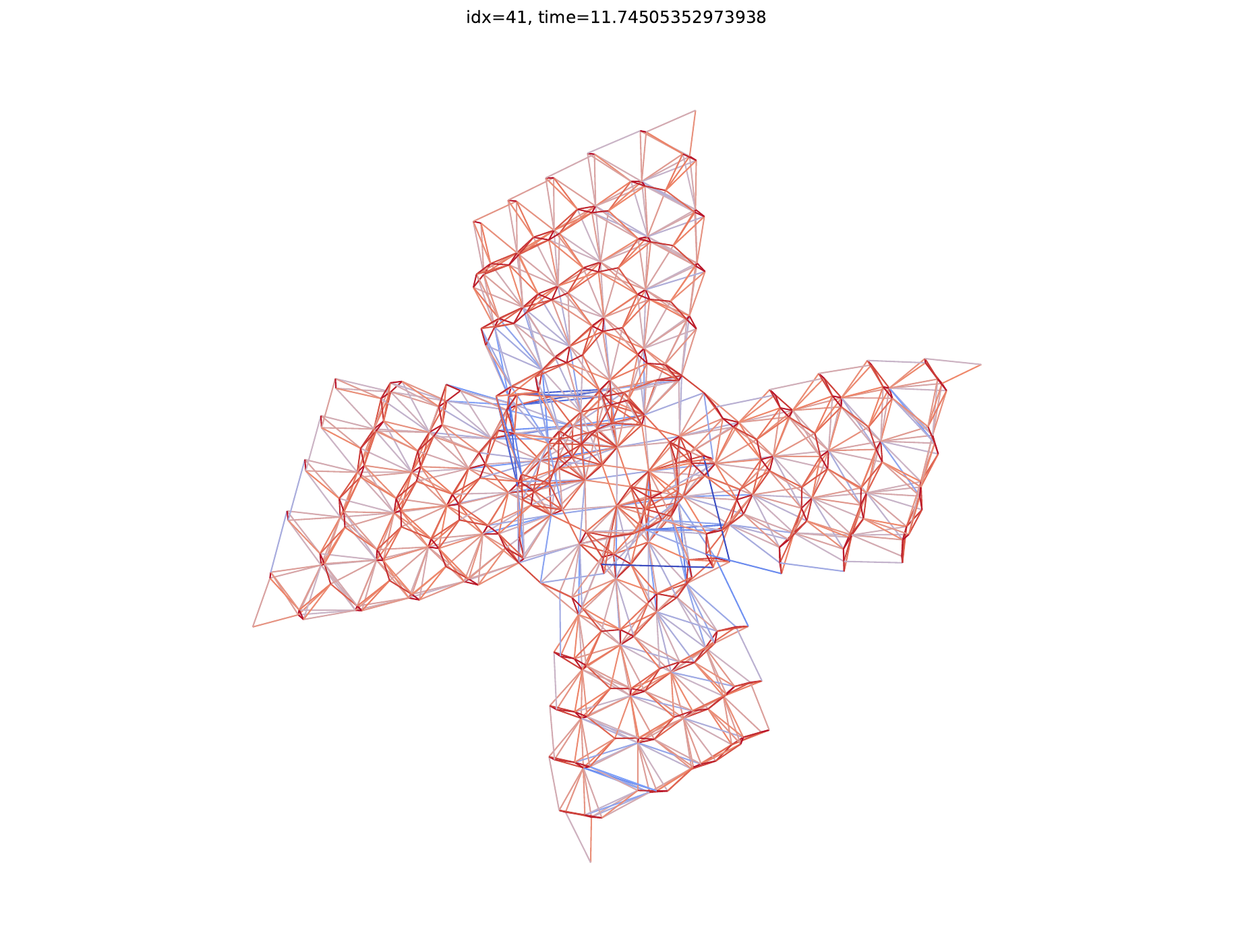} &
\imgcell{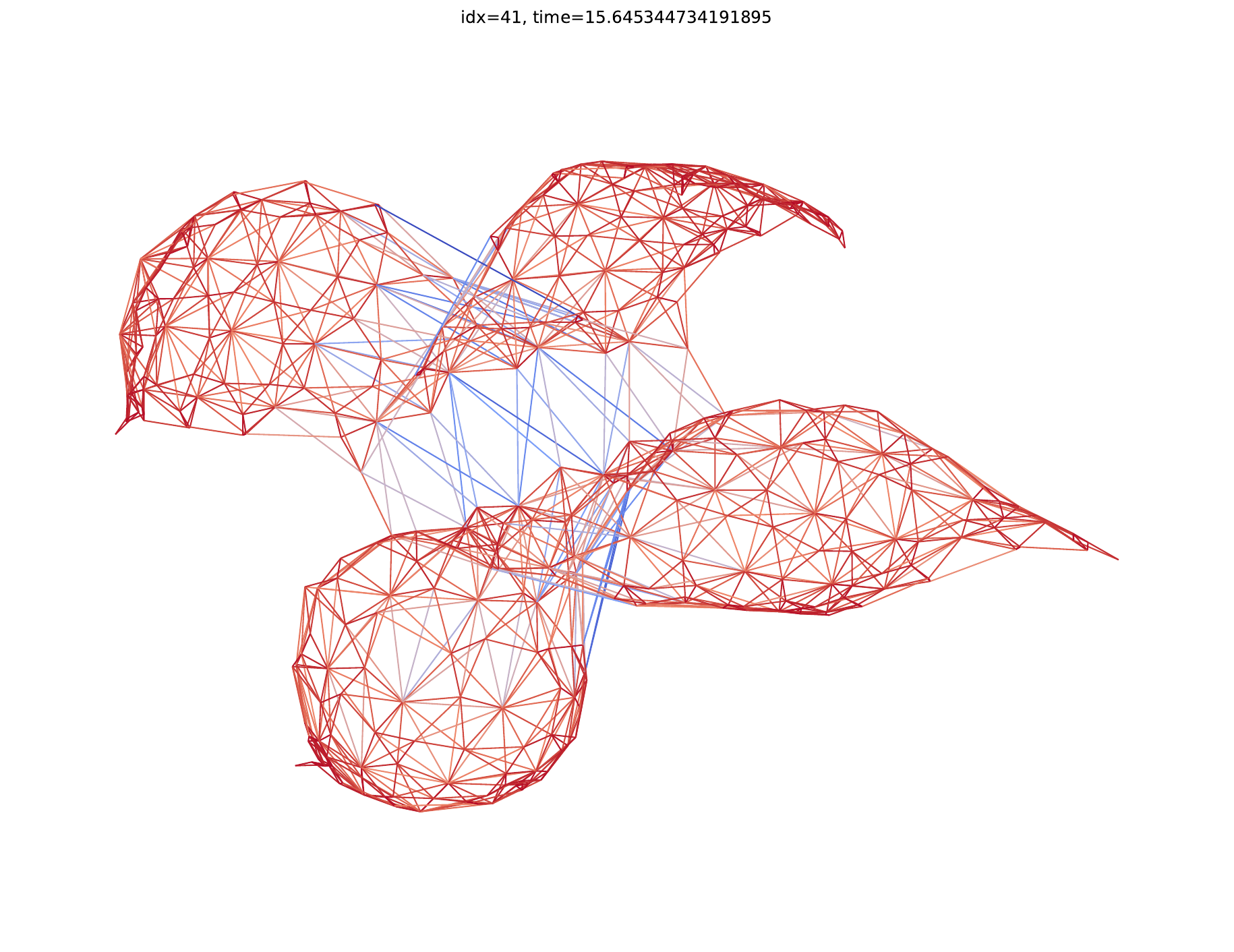} &
\imgcell{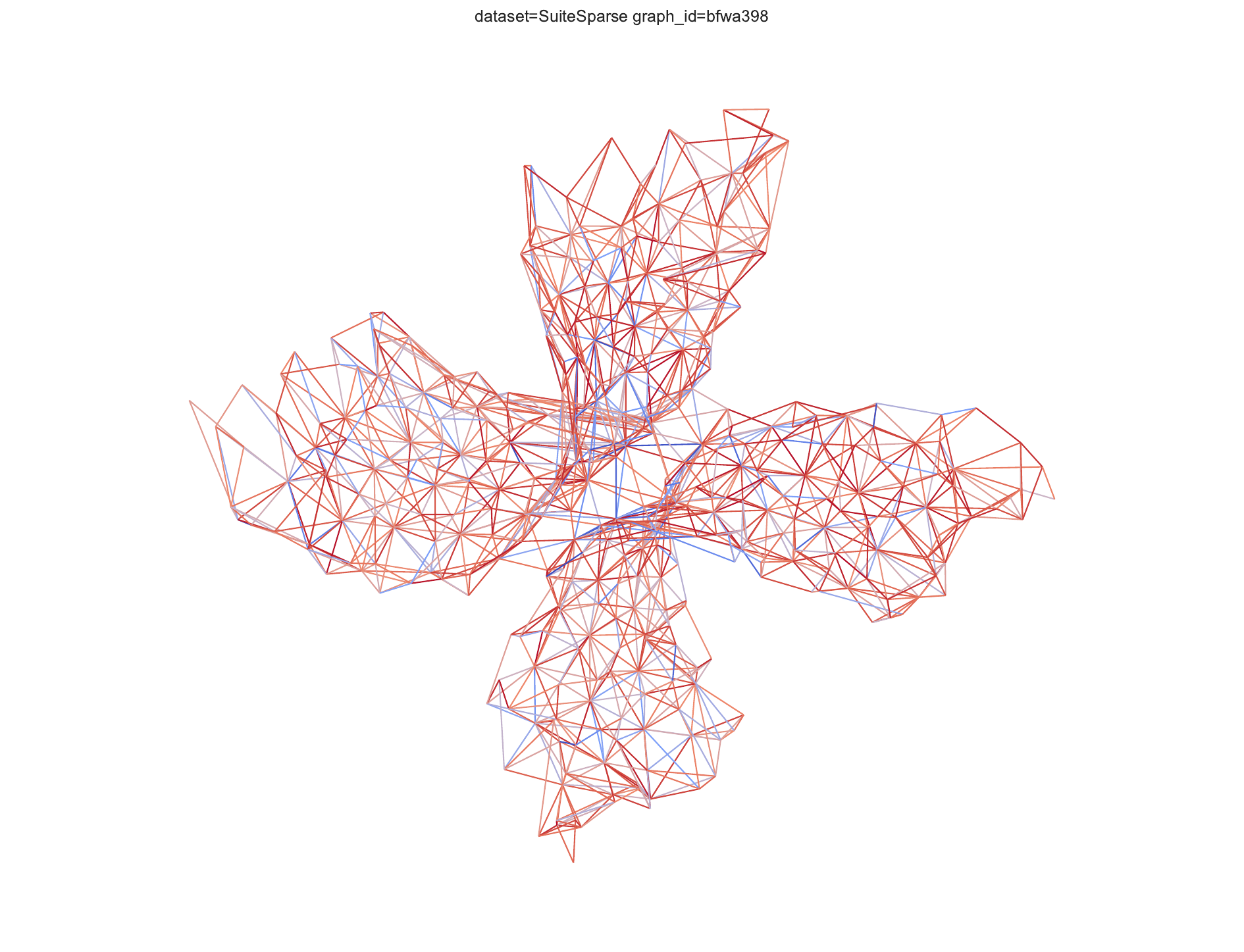} &
\imgcell{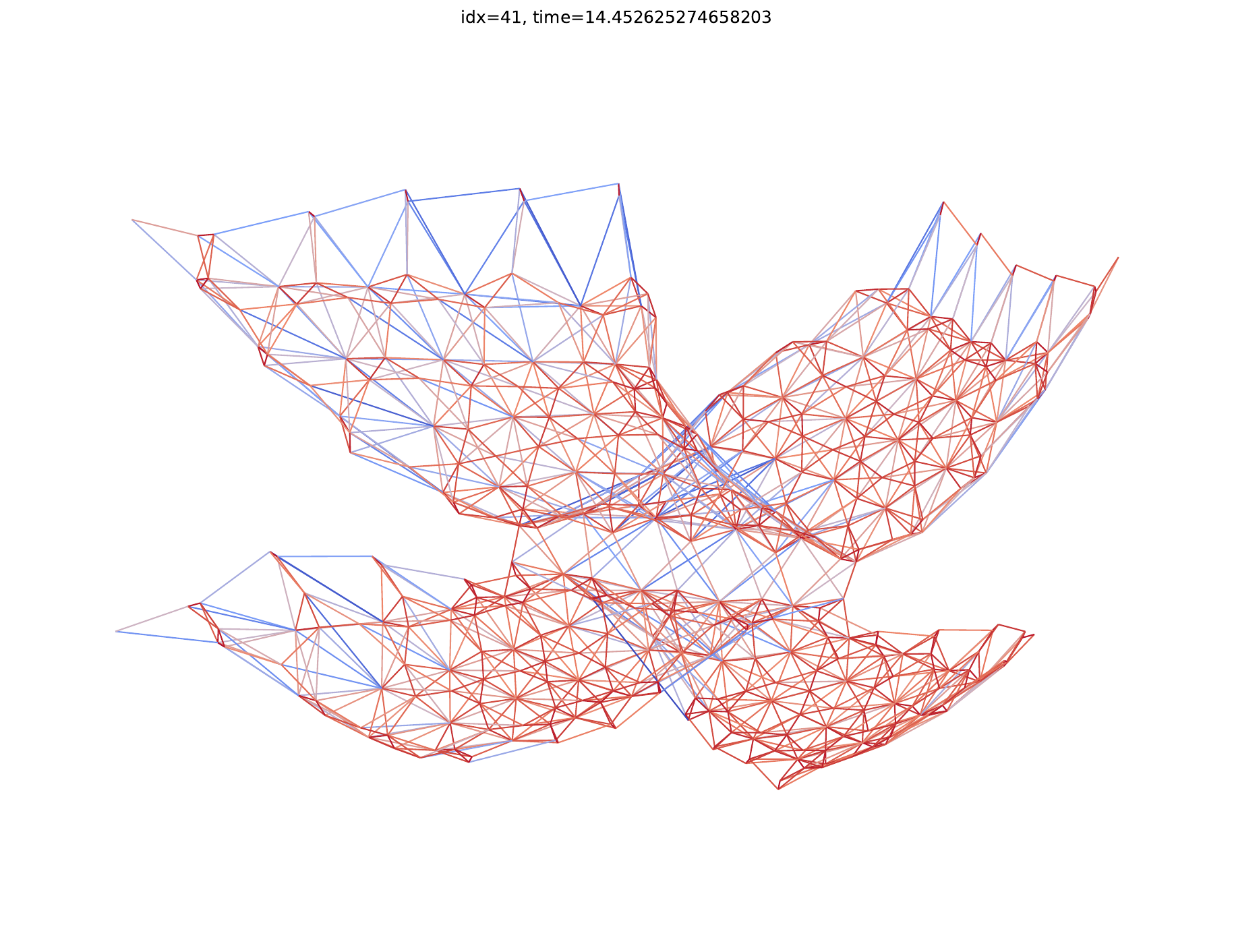} &
\imgcell{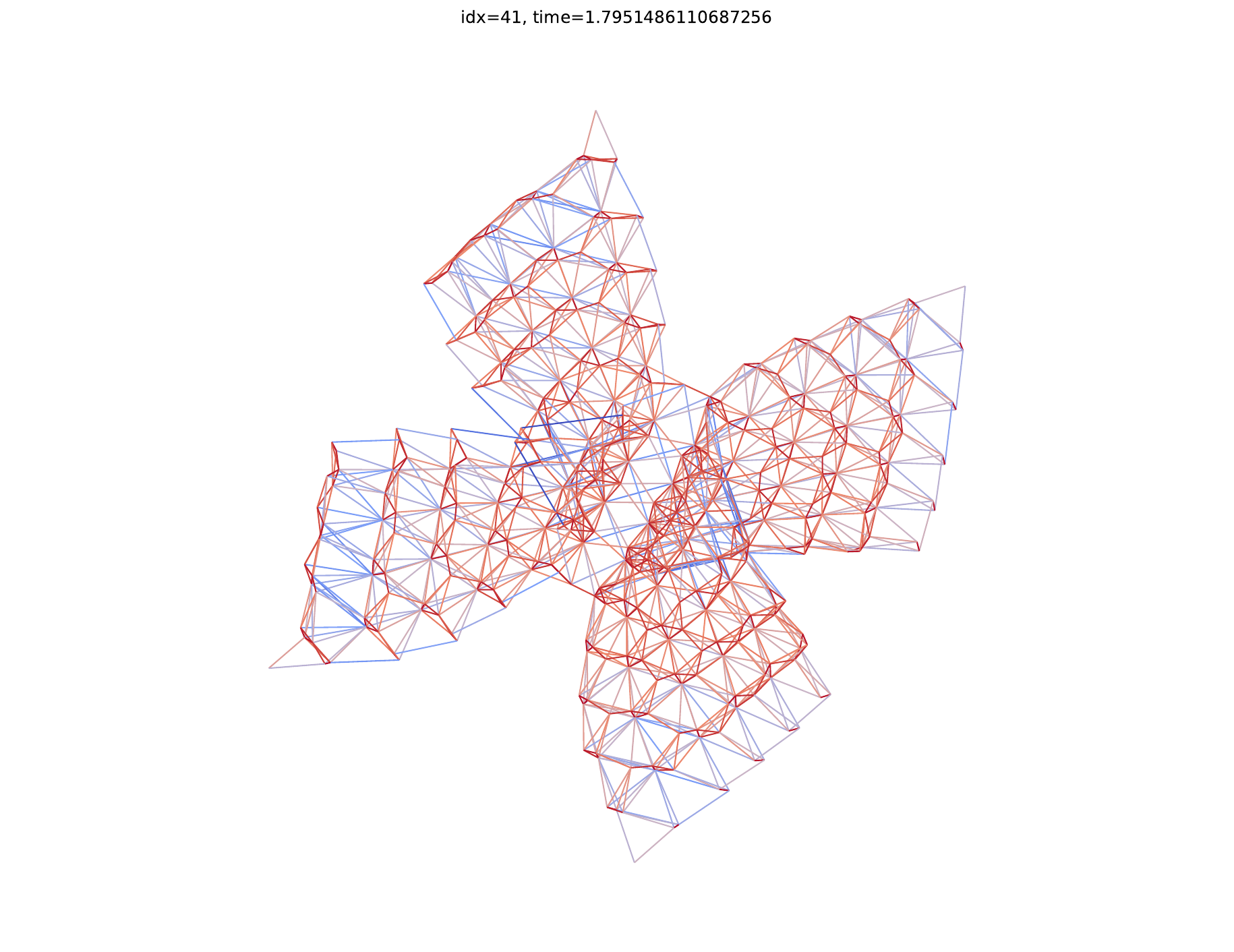} &
\imgcell{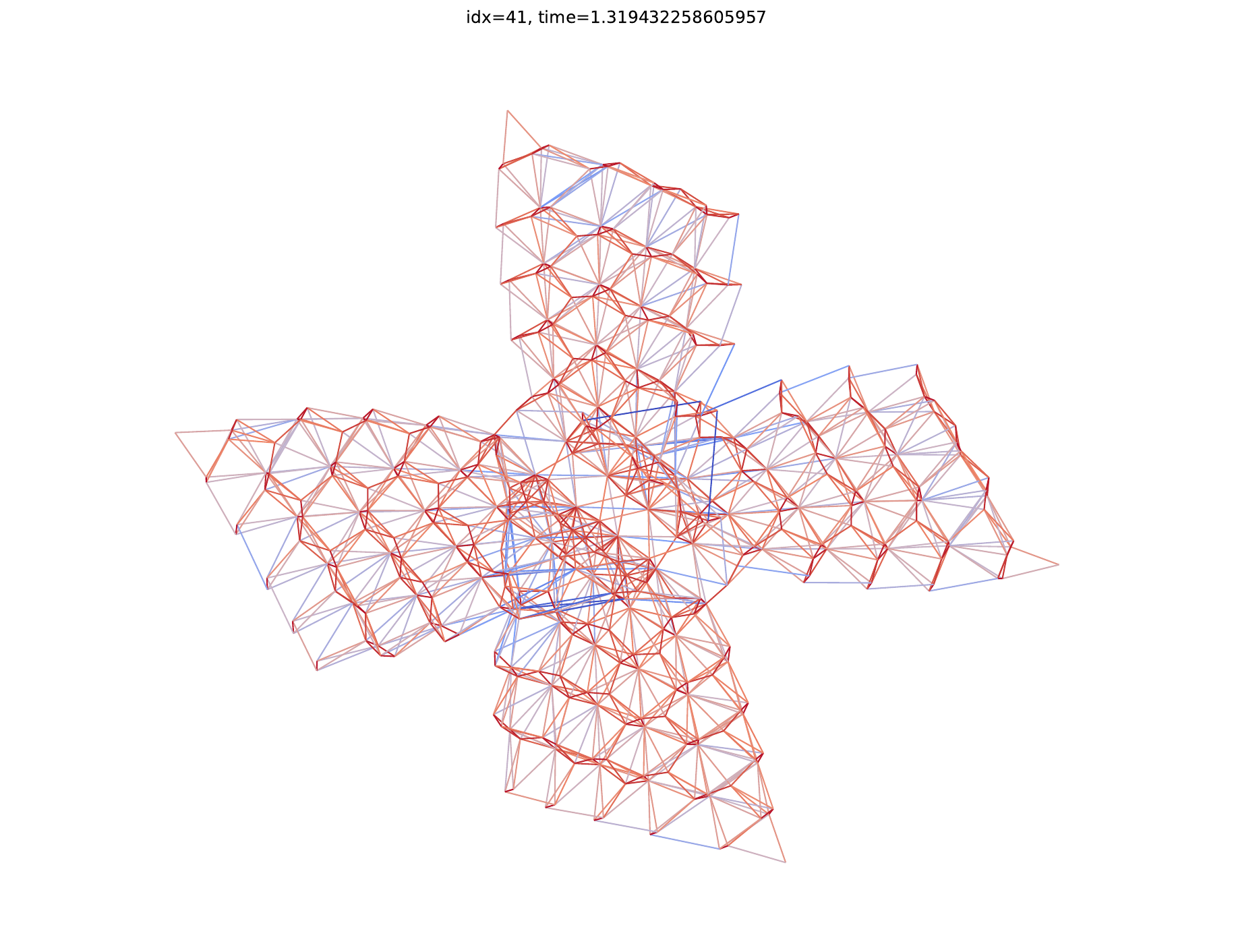} &
\imgcell{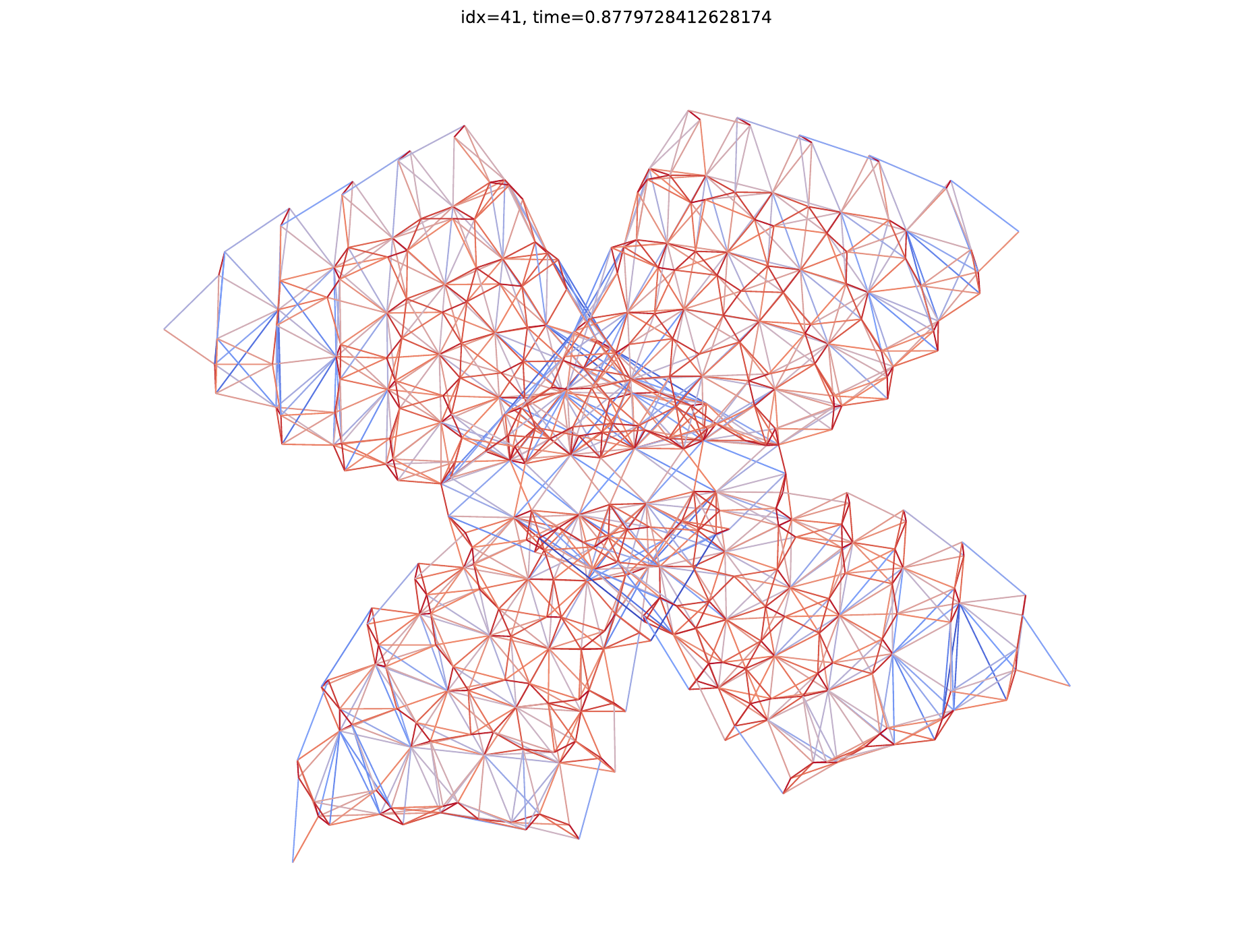} \\

&
t = 0.05s &
t = 8.25s &
t = 12.10s &
t = 0.75s &
t = 7200.00s &
t = 0.61s &
t = 0.85s &
t = 0.76s &
t = 0.72s &
t = 0.81s &
t = 0.65s &
t = 0.58s \\

\makecell{\bfseries hor\_131\\N = 434\\M = 2138} &
\imgcell{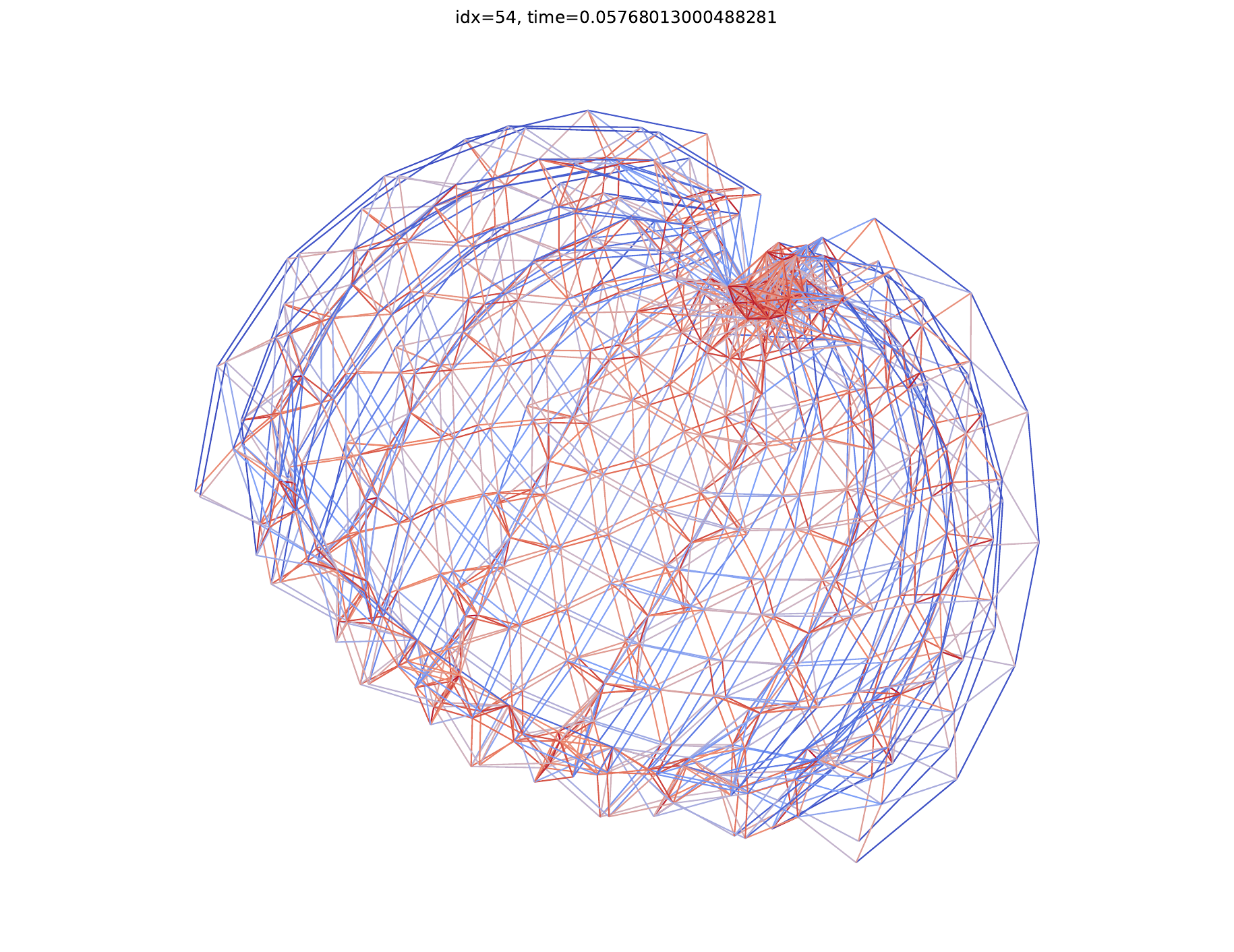} &
\imgcell{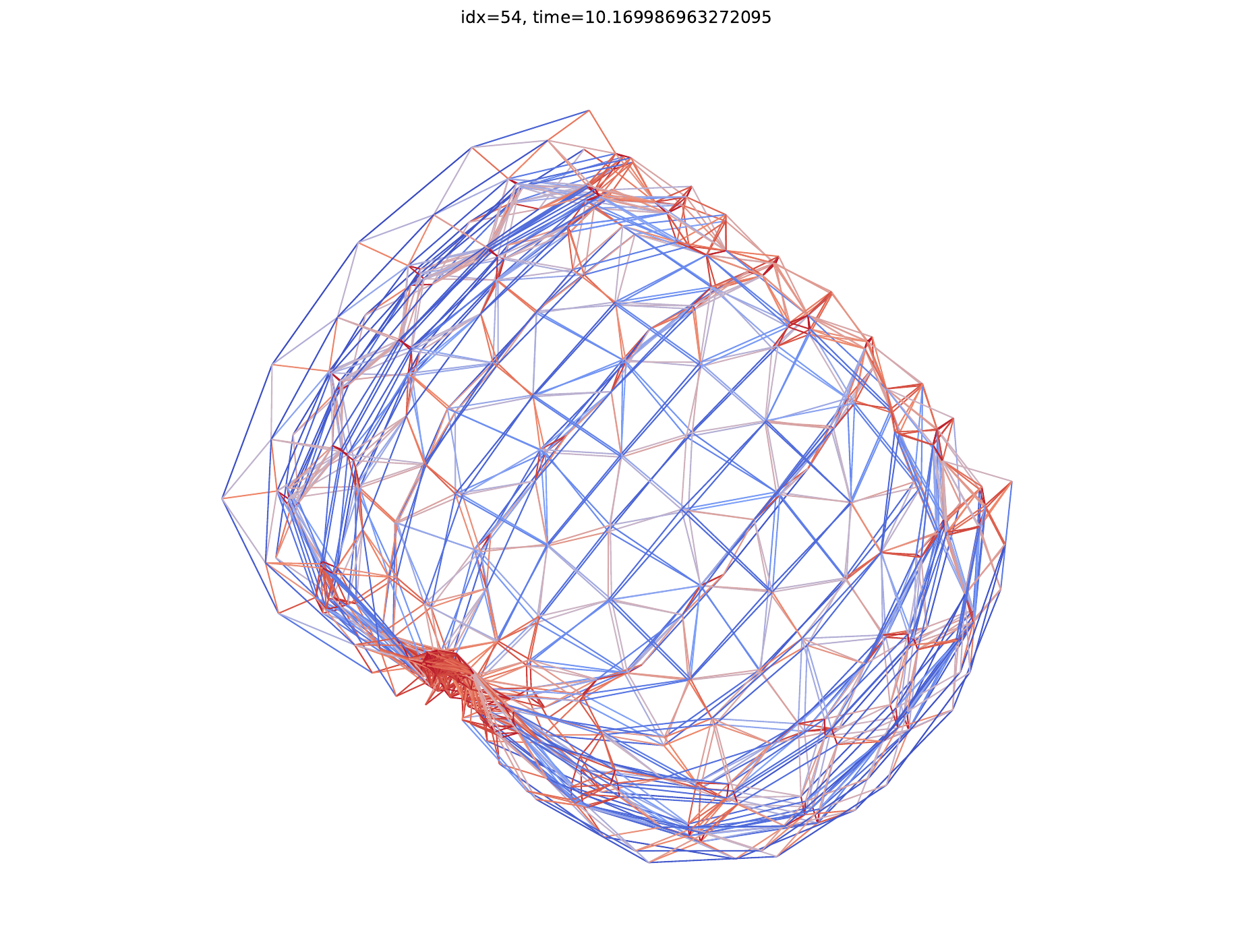} &
\imgcell{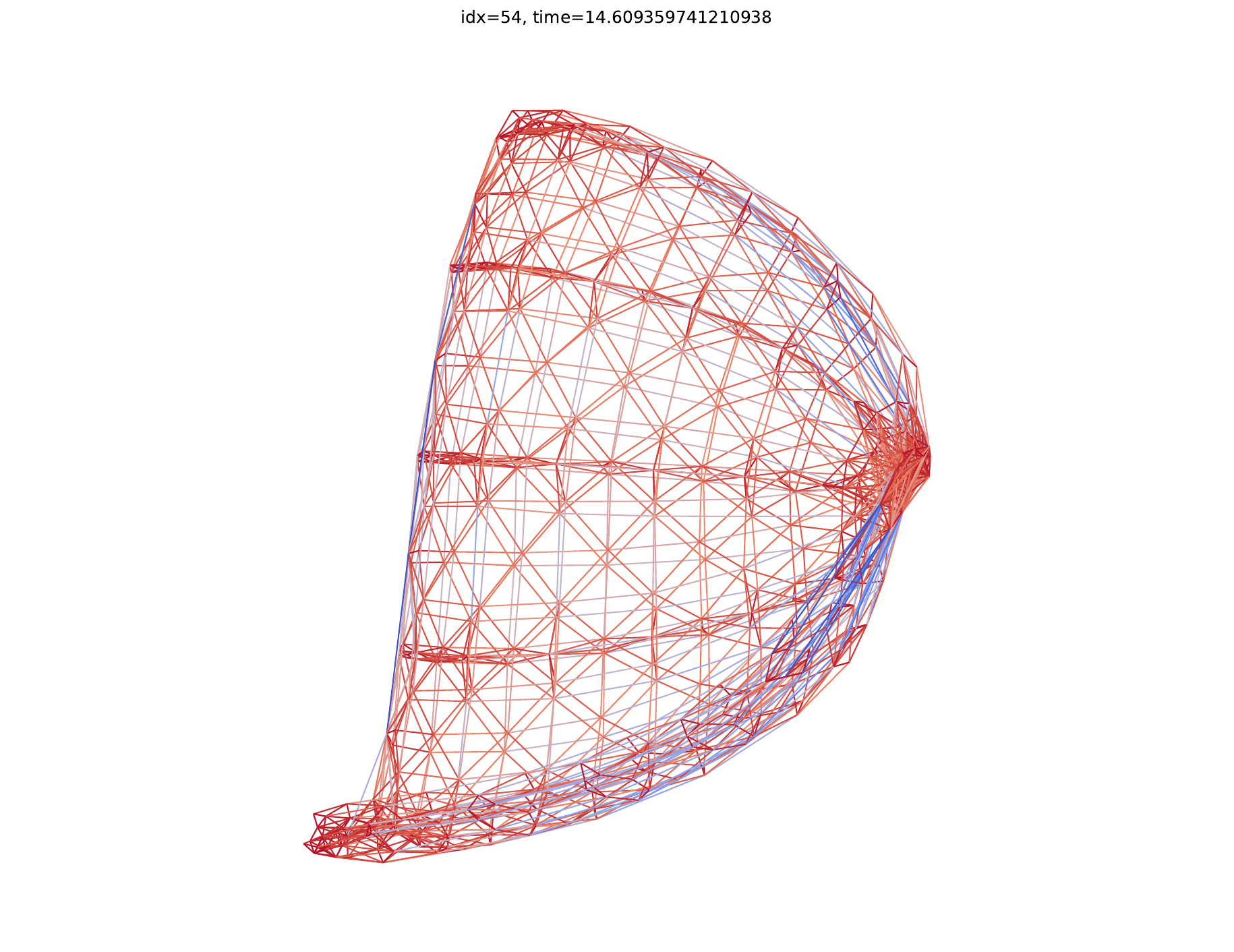} &
\imgcell{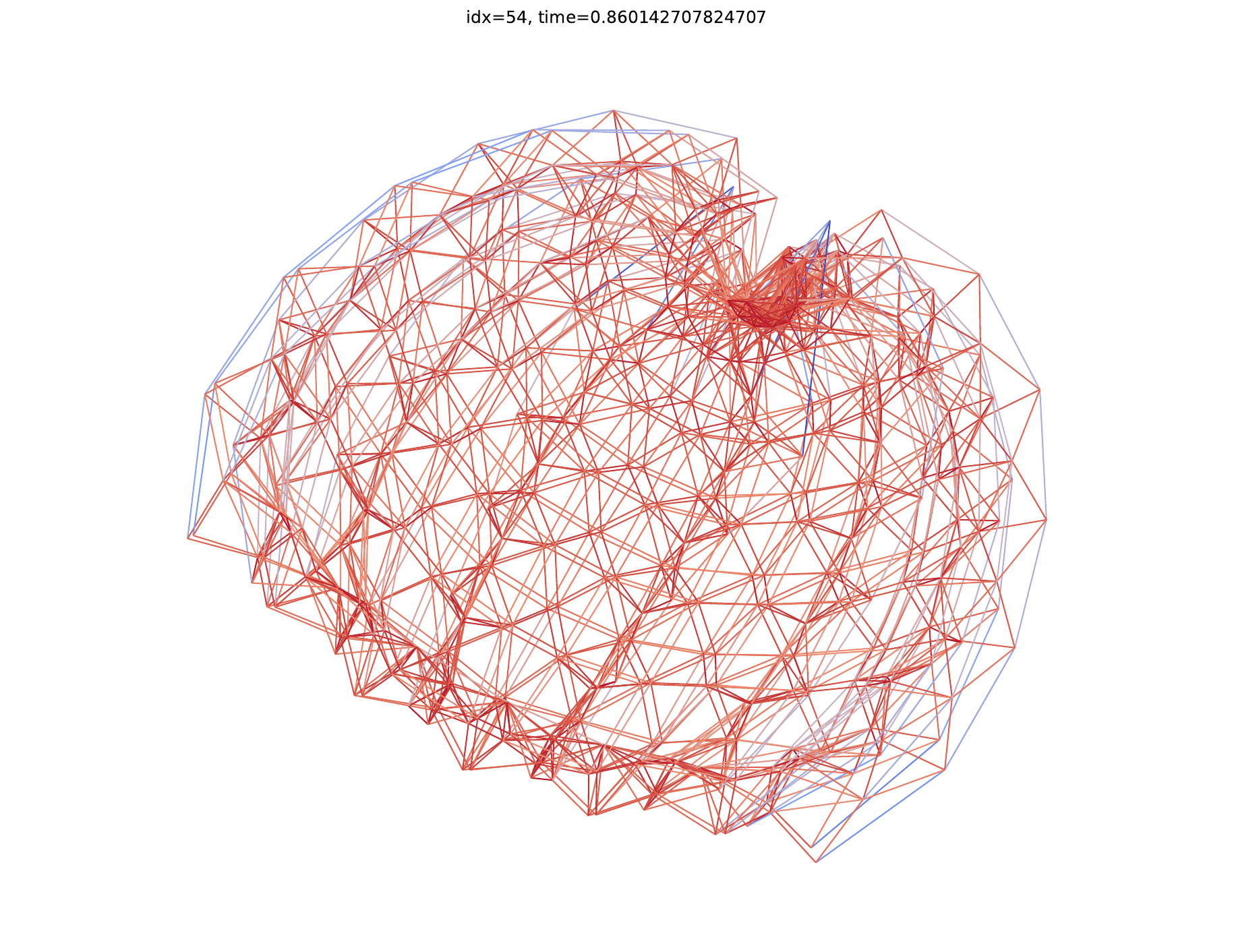} &
\imgcell{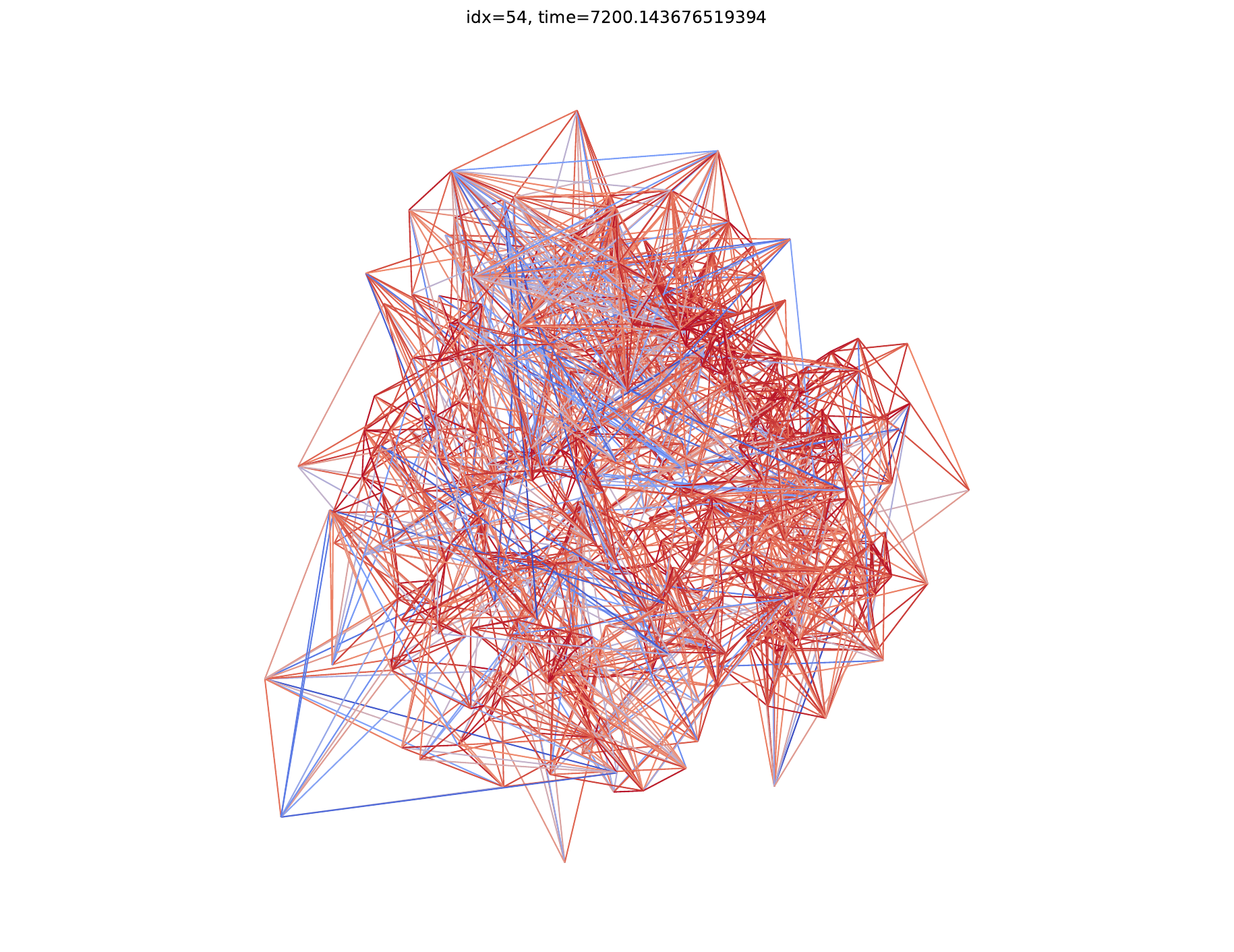} &
\imgcell{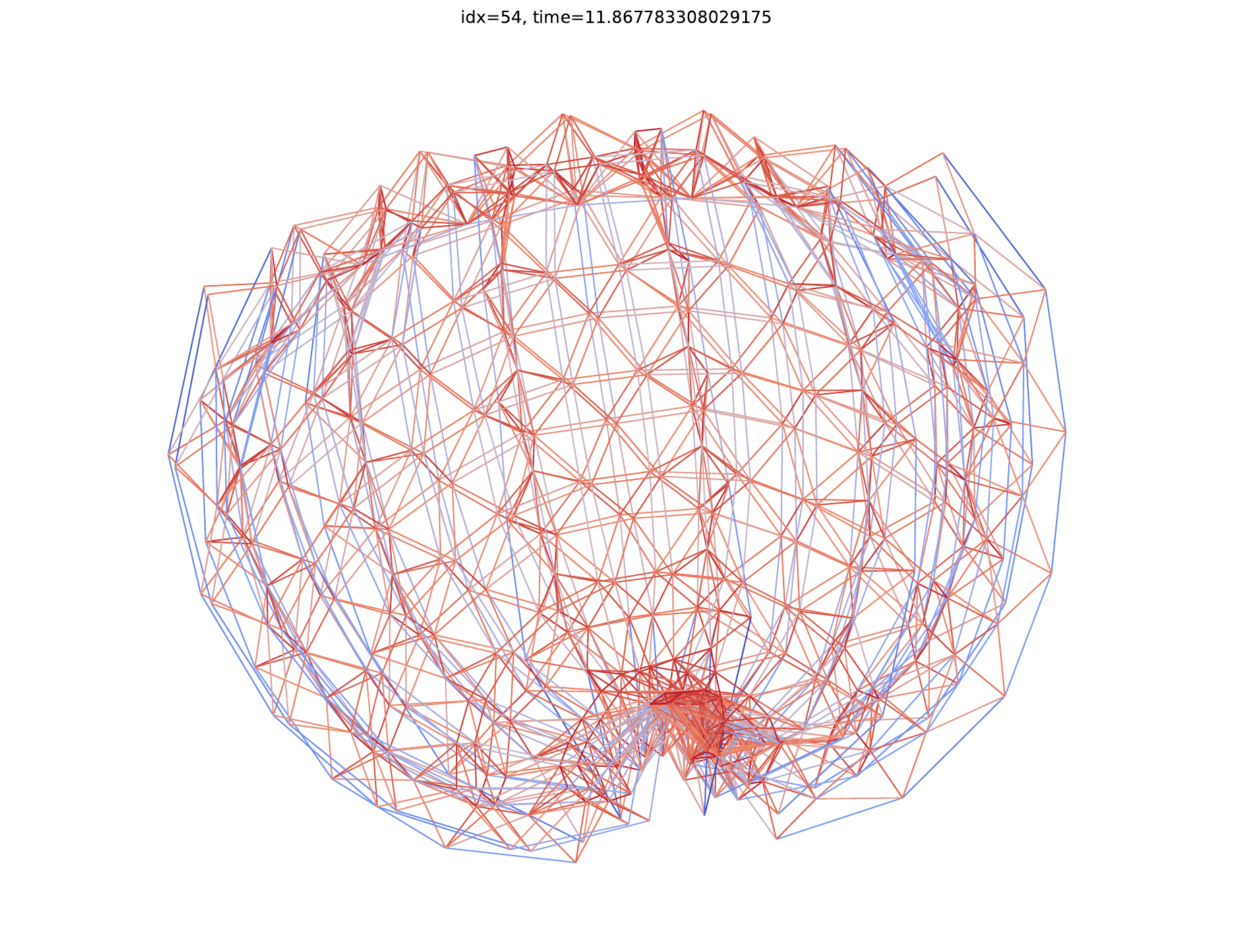} &
\imgcell{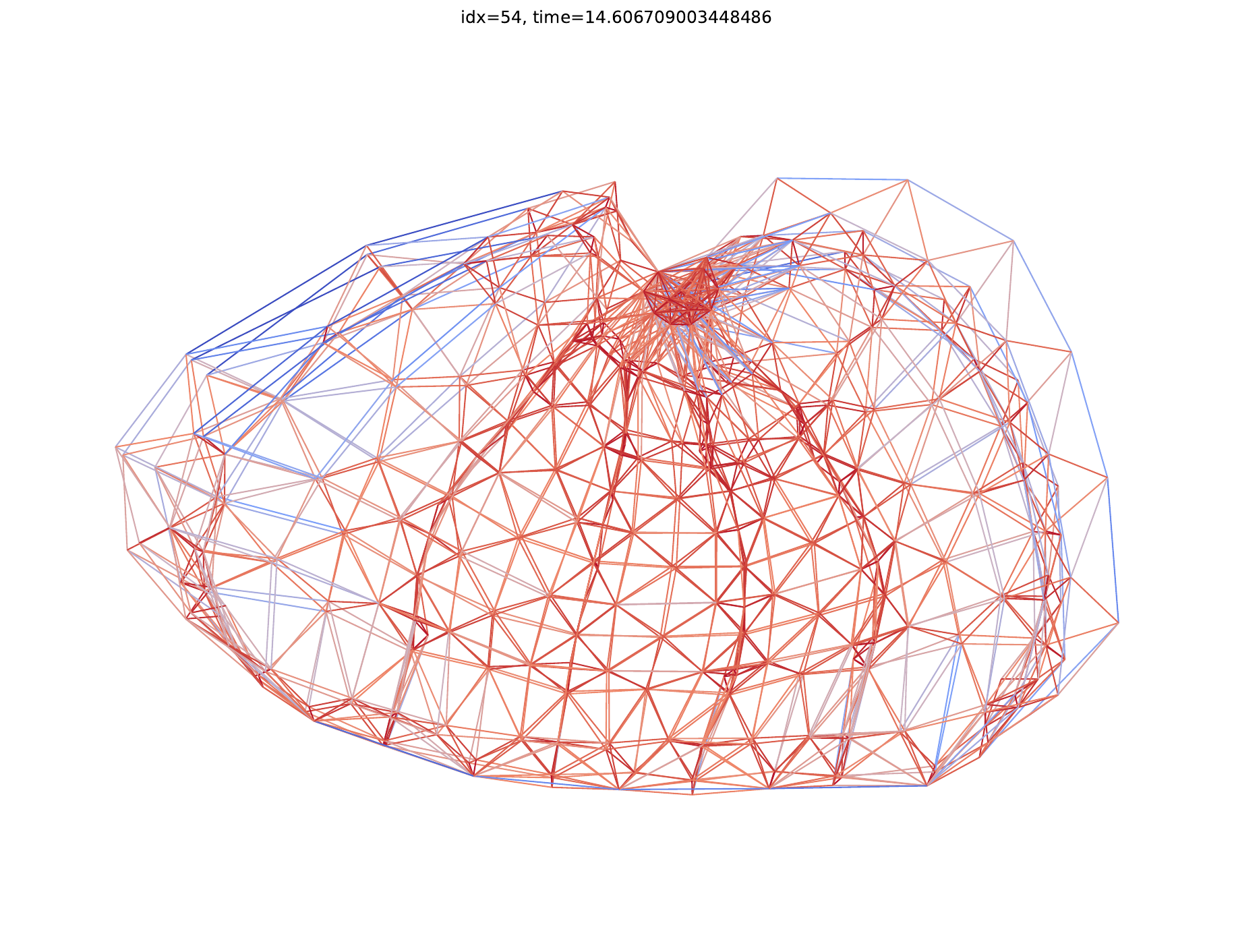} &
\imgcell{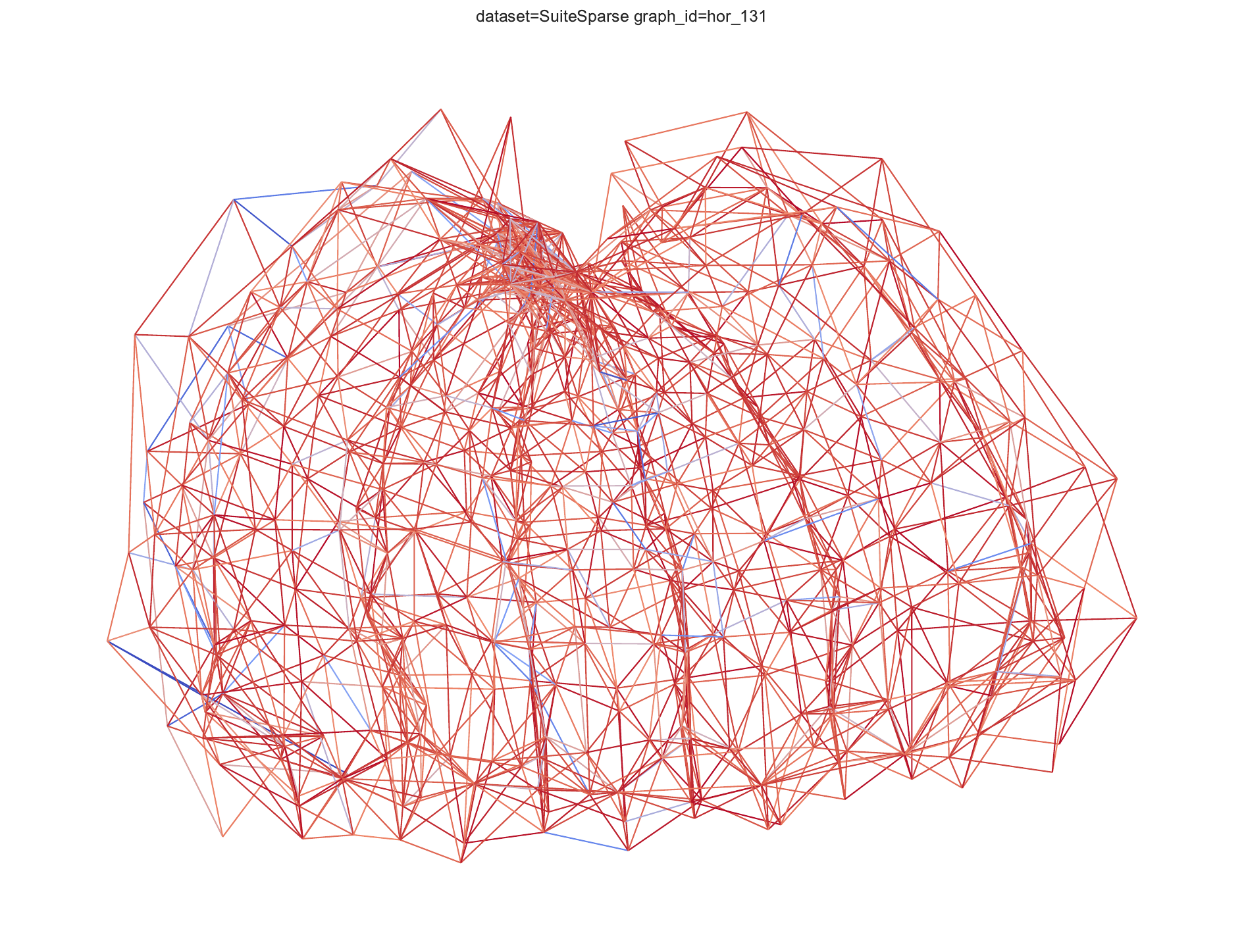} &
\imgcell{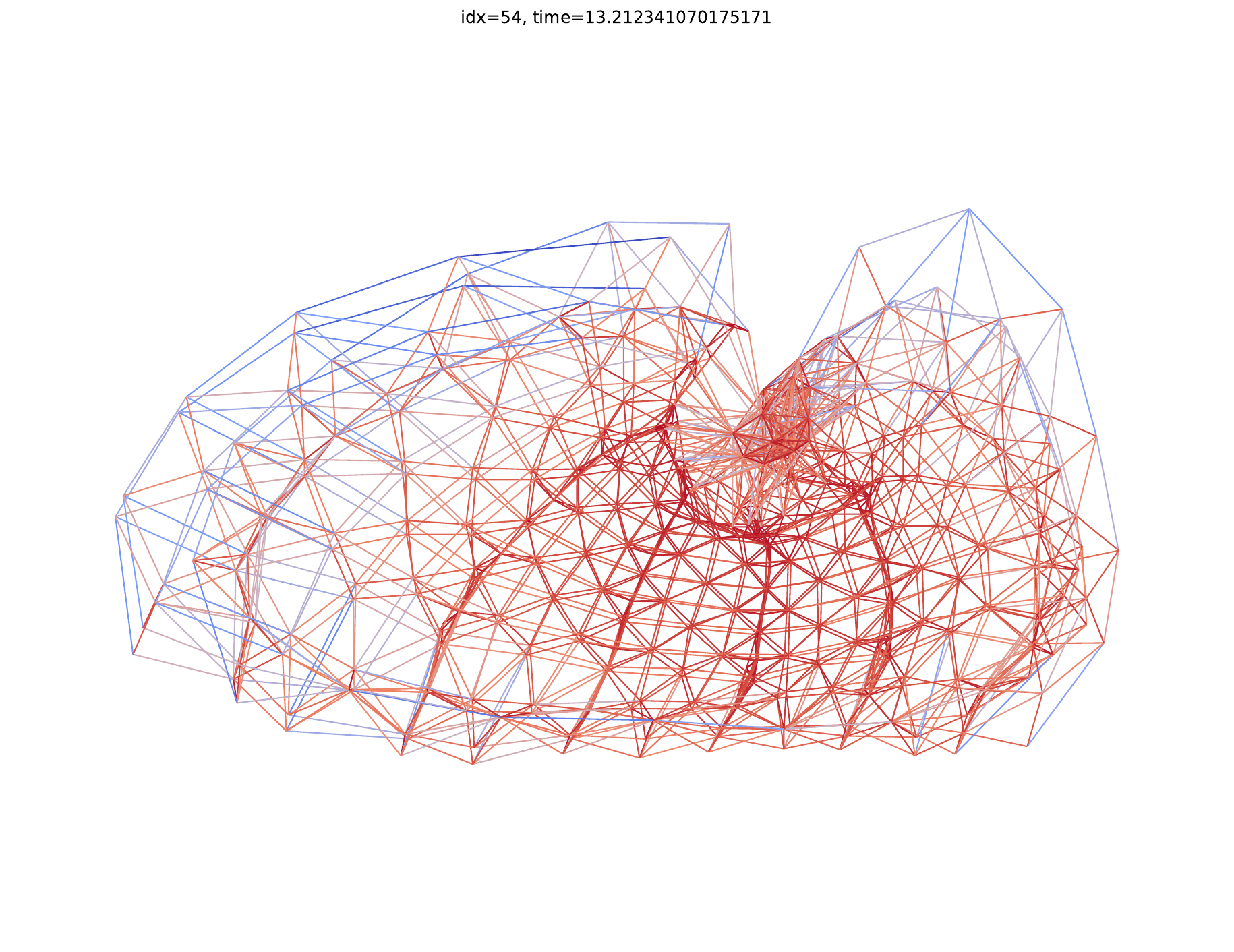} &
\imgcell{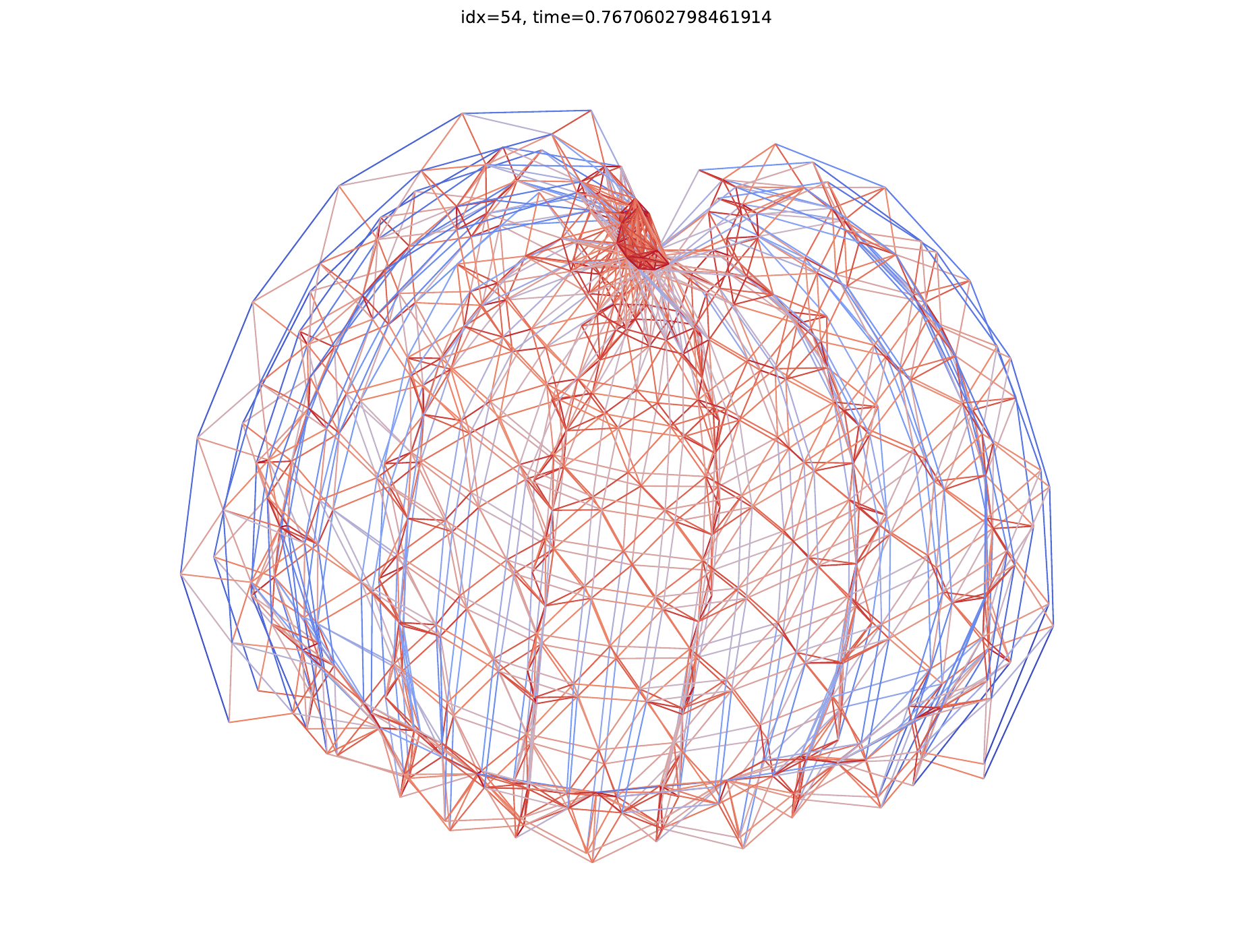} &
\imgcell{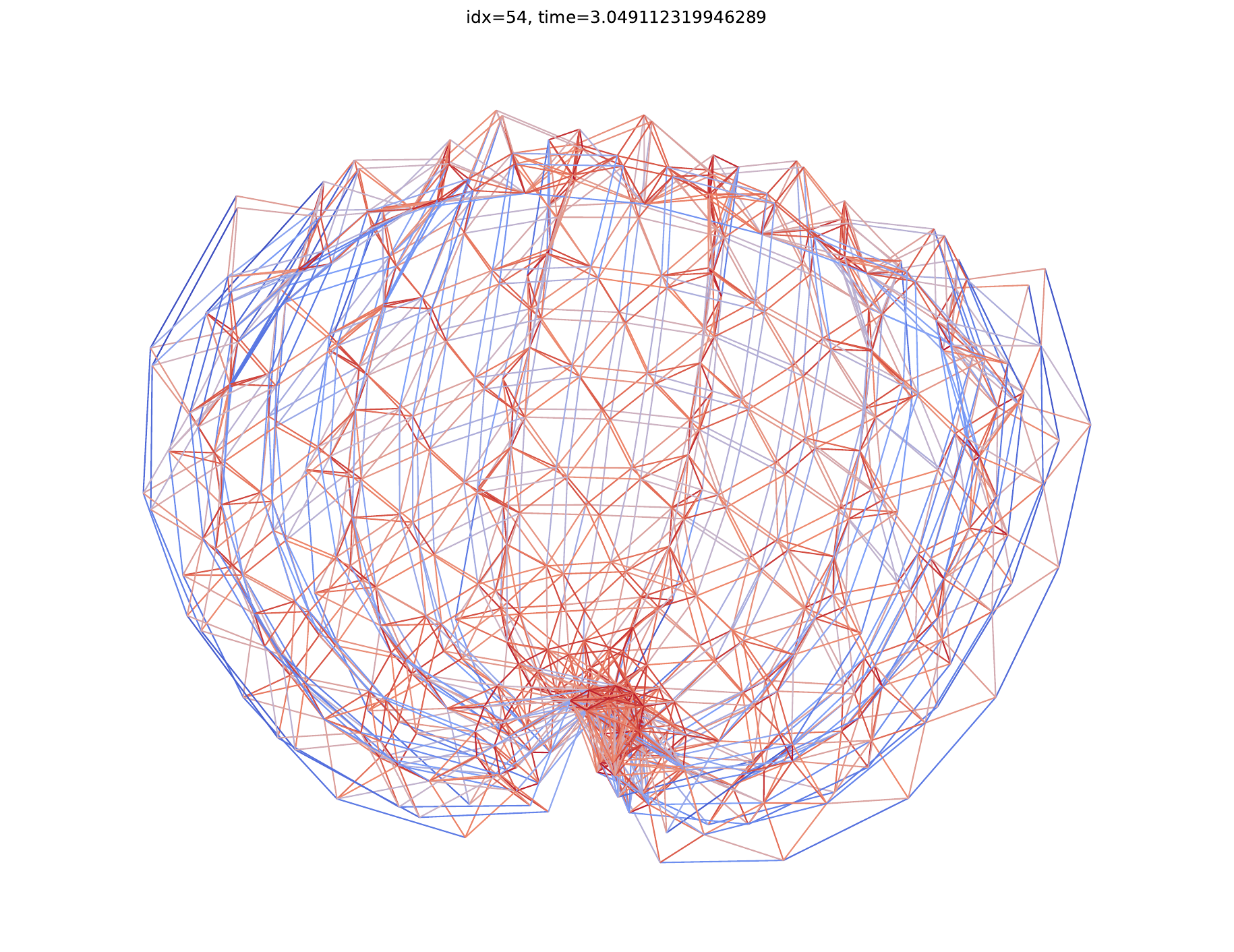} &
\imgcell{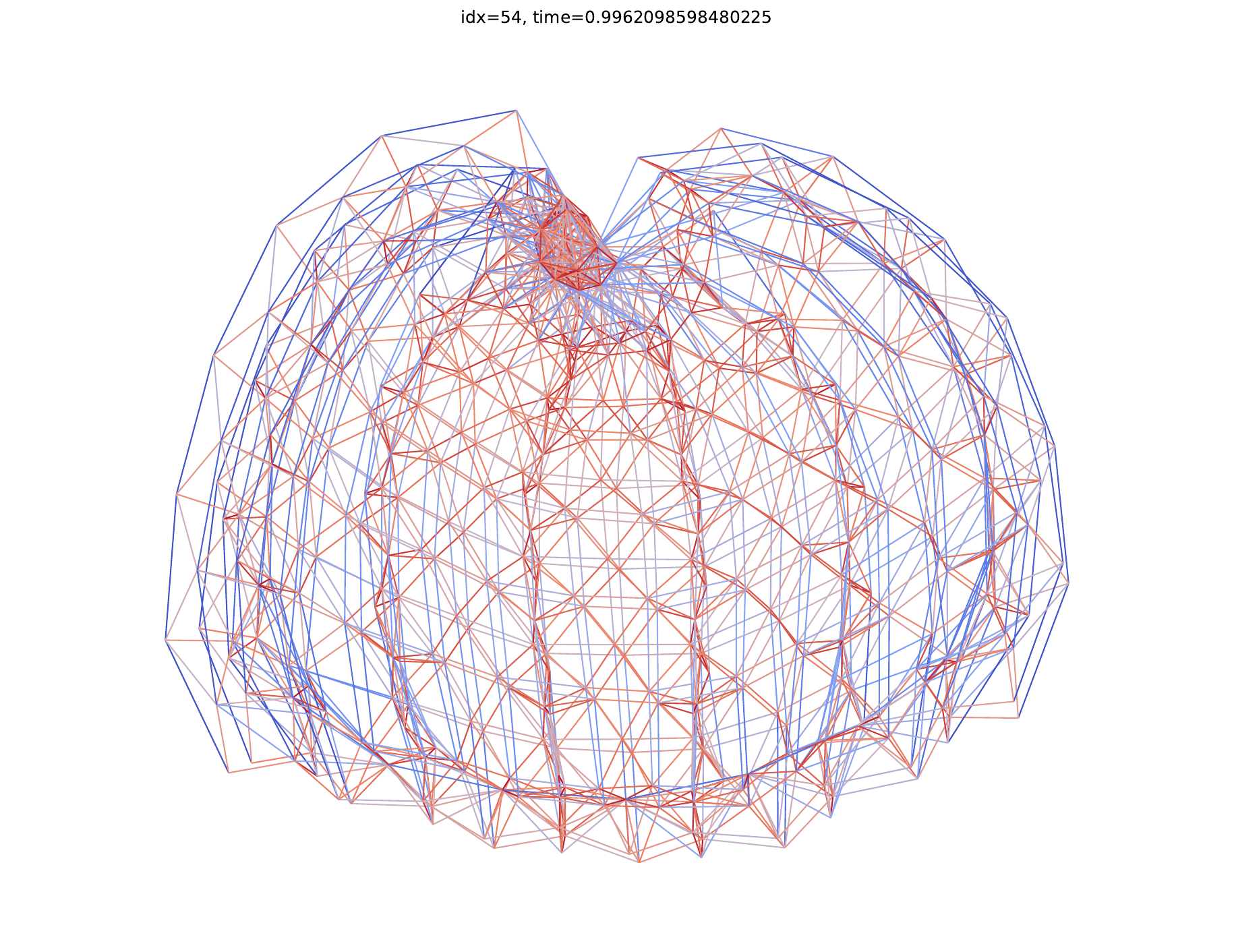} \\

&
t = 0.06s &
t = 10.17s &
t = 14.61s &
t = 0.86s &
t = 7200.00s &
t = 0.88s &
t = 0.76s &
t = 0.73s &
t = 0.65s &
t = 0.75s &
t = 0.63s &
t = 0.66s \\

\makecell{\bfseries can\_445\\N = 445\\M = 1682} &
\imgcell{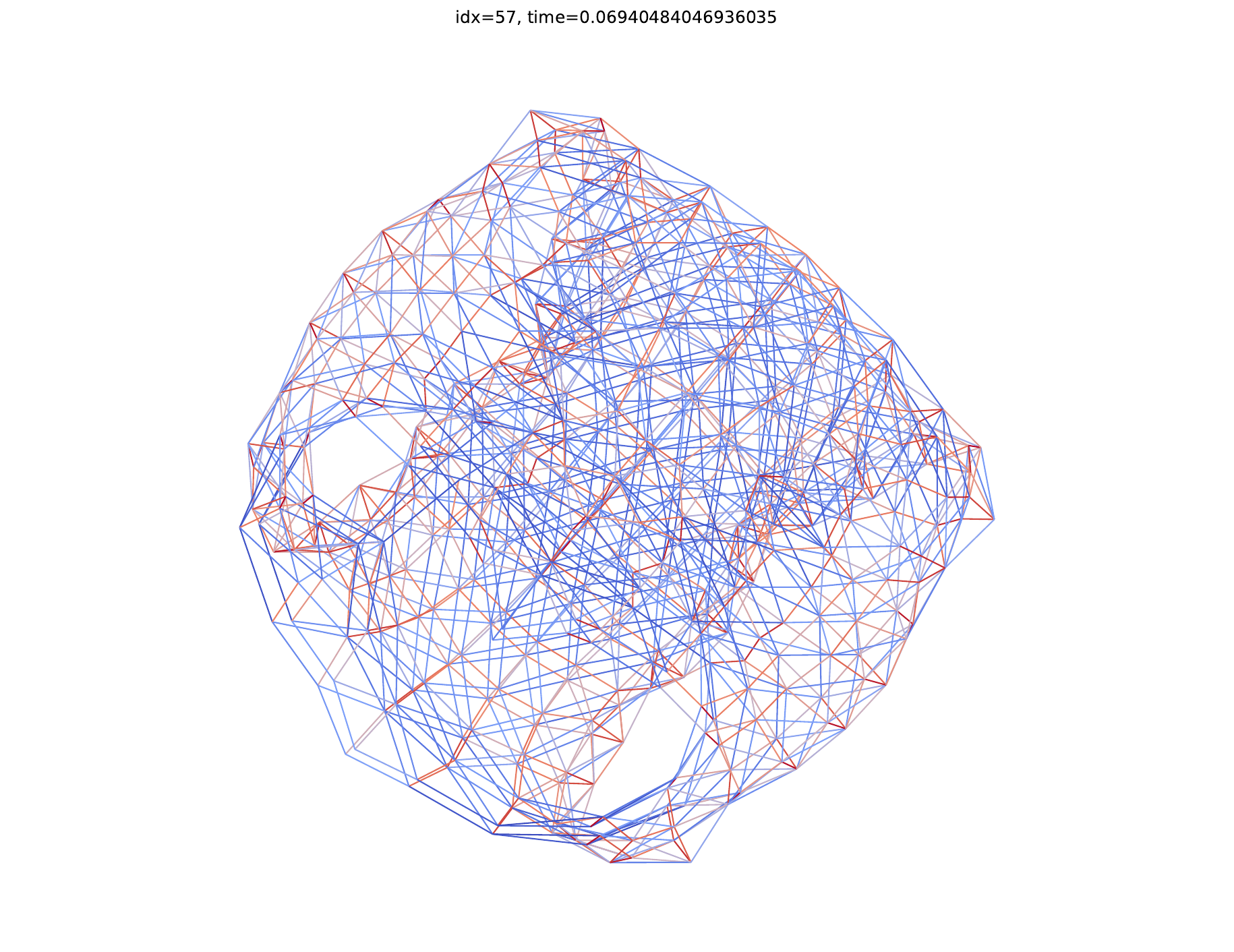} &
\imgcell{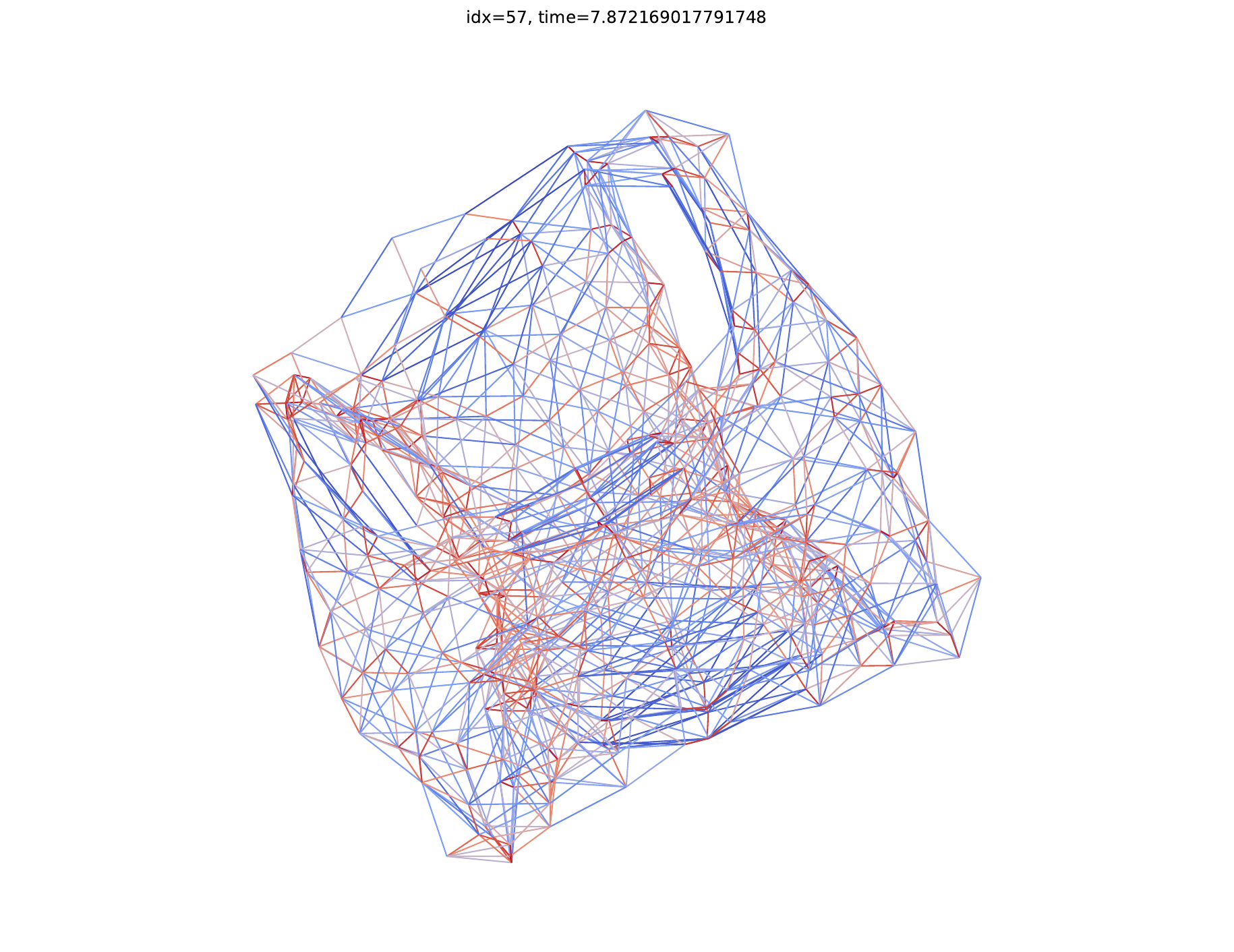} &
\imgcell{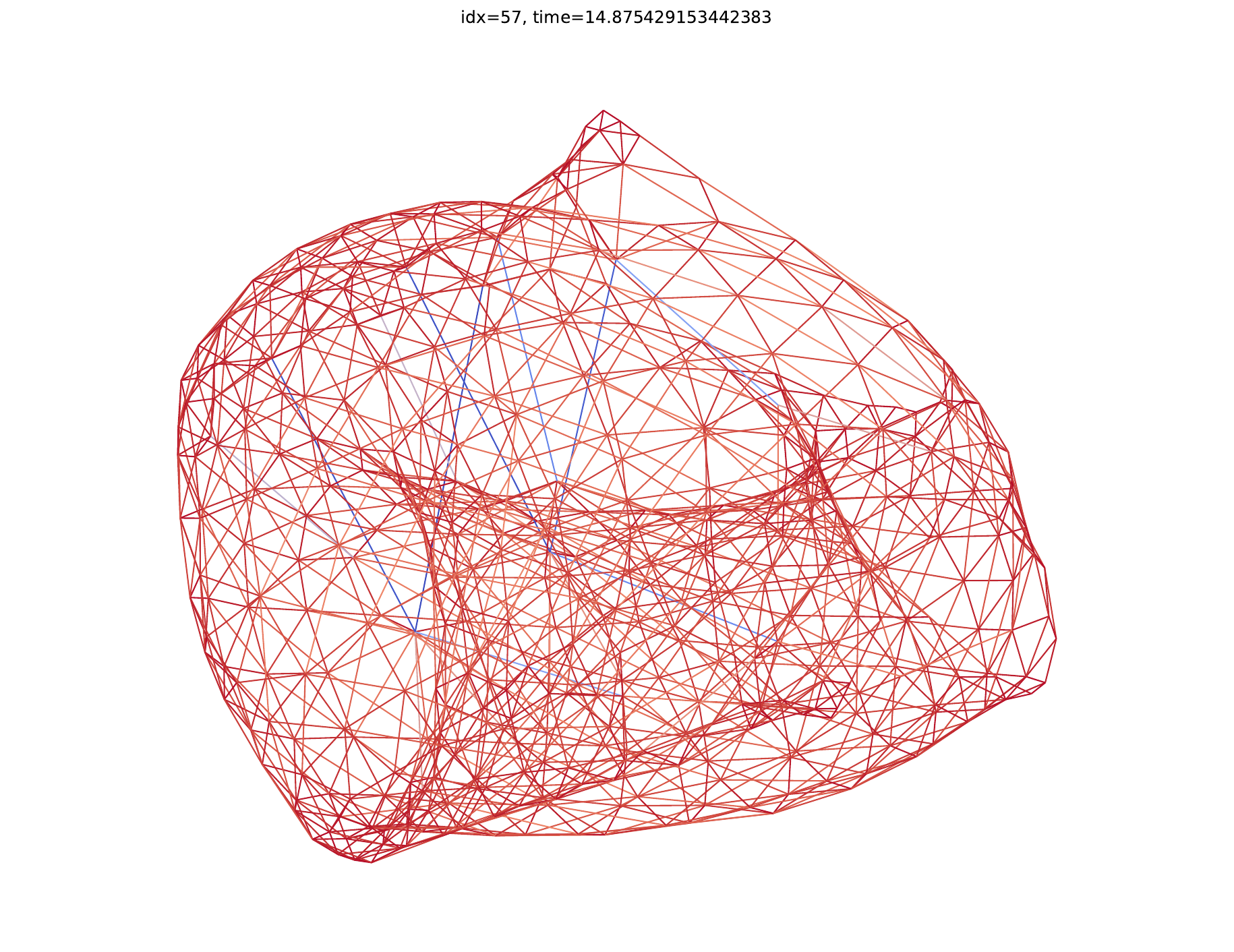} &
\imgcell{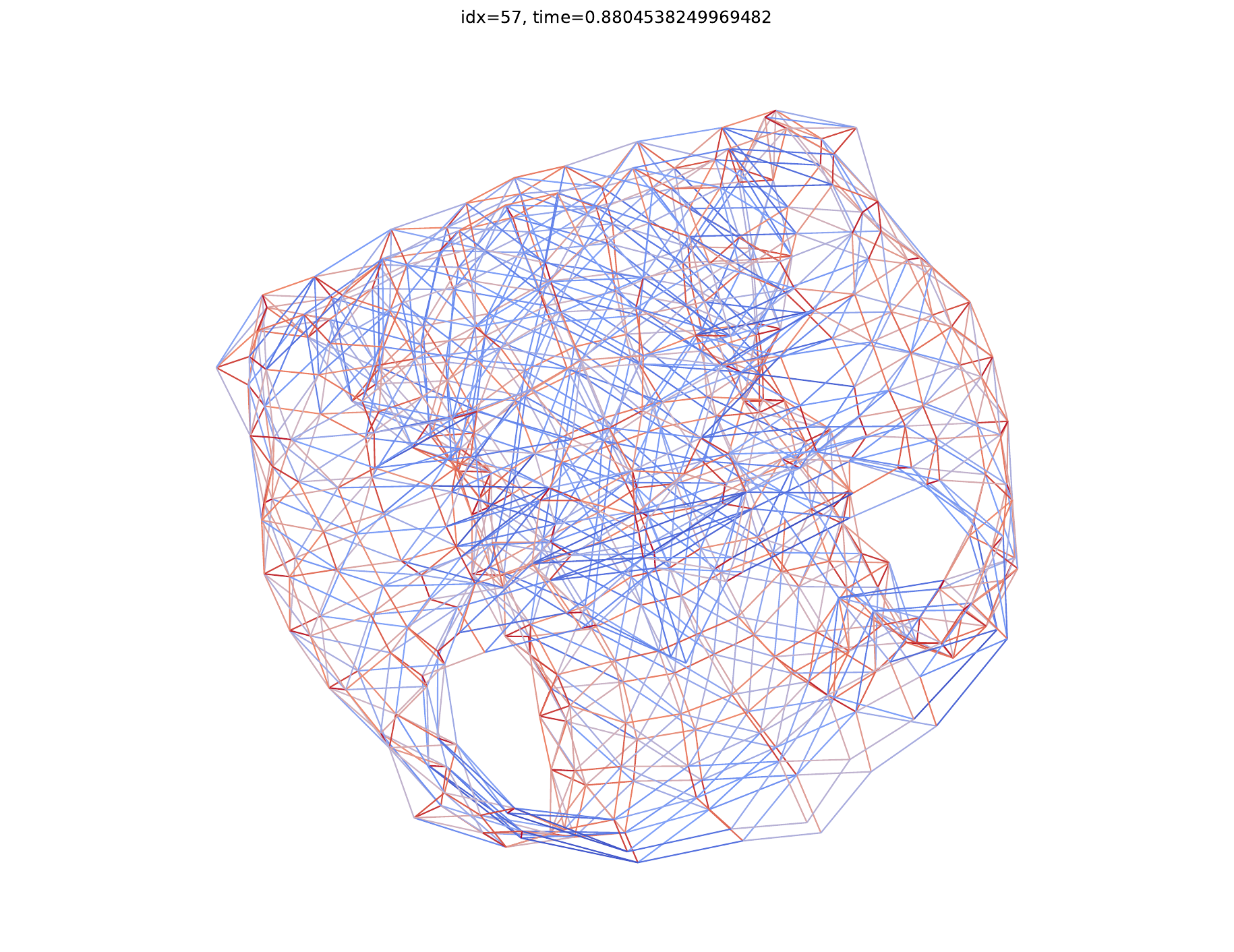} &
\imgcell{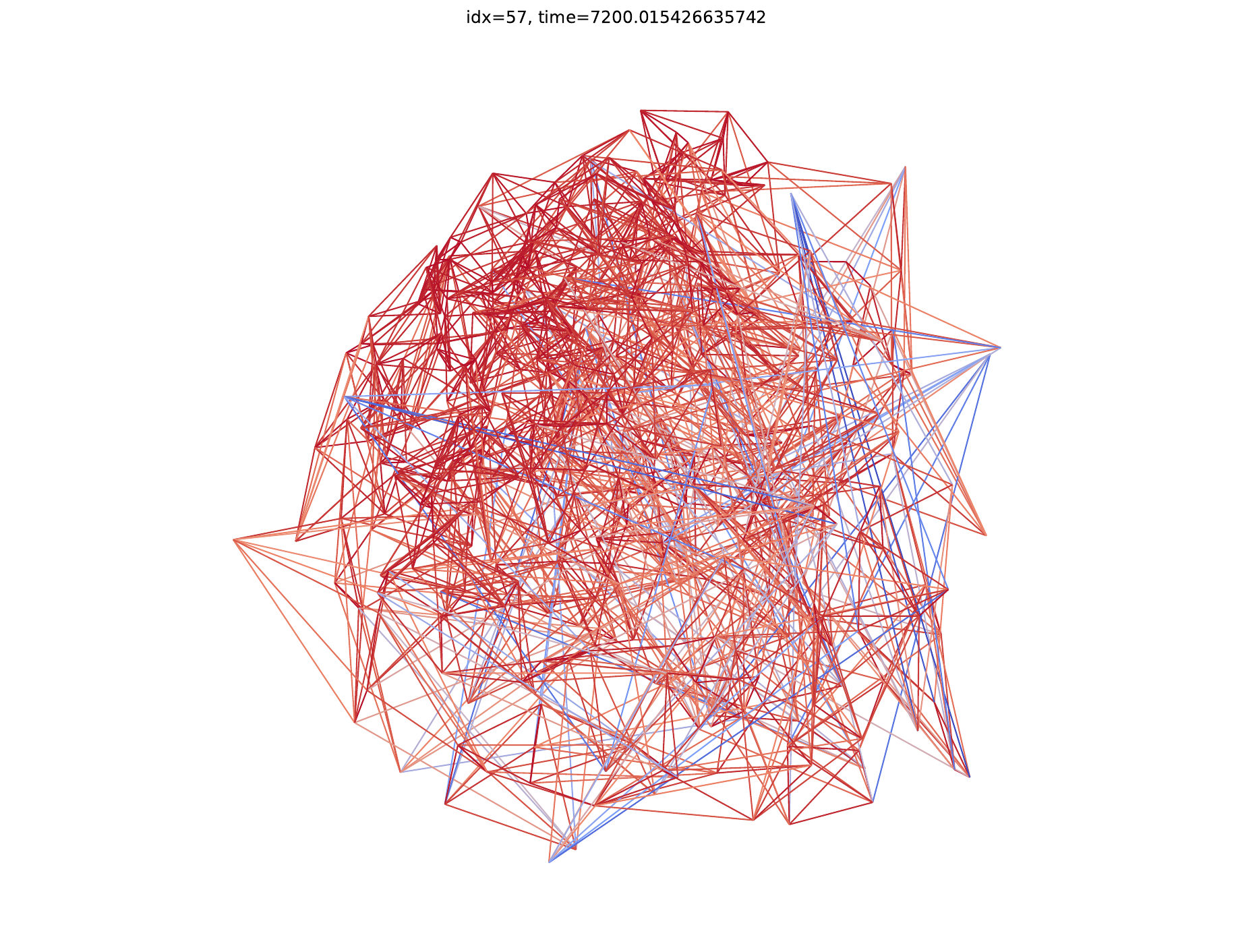} &
\imgcell{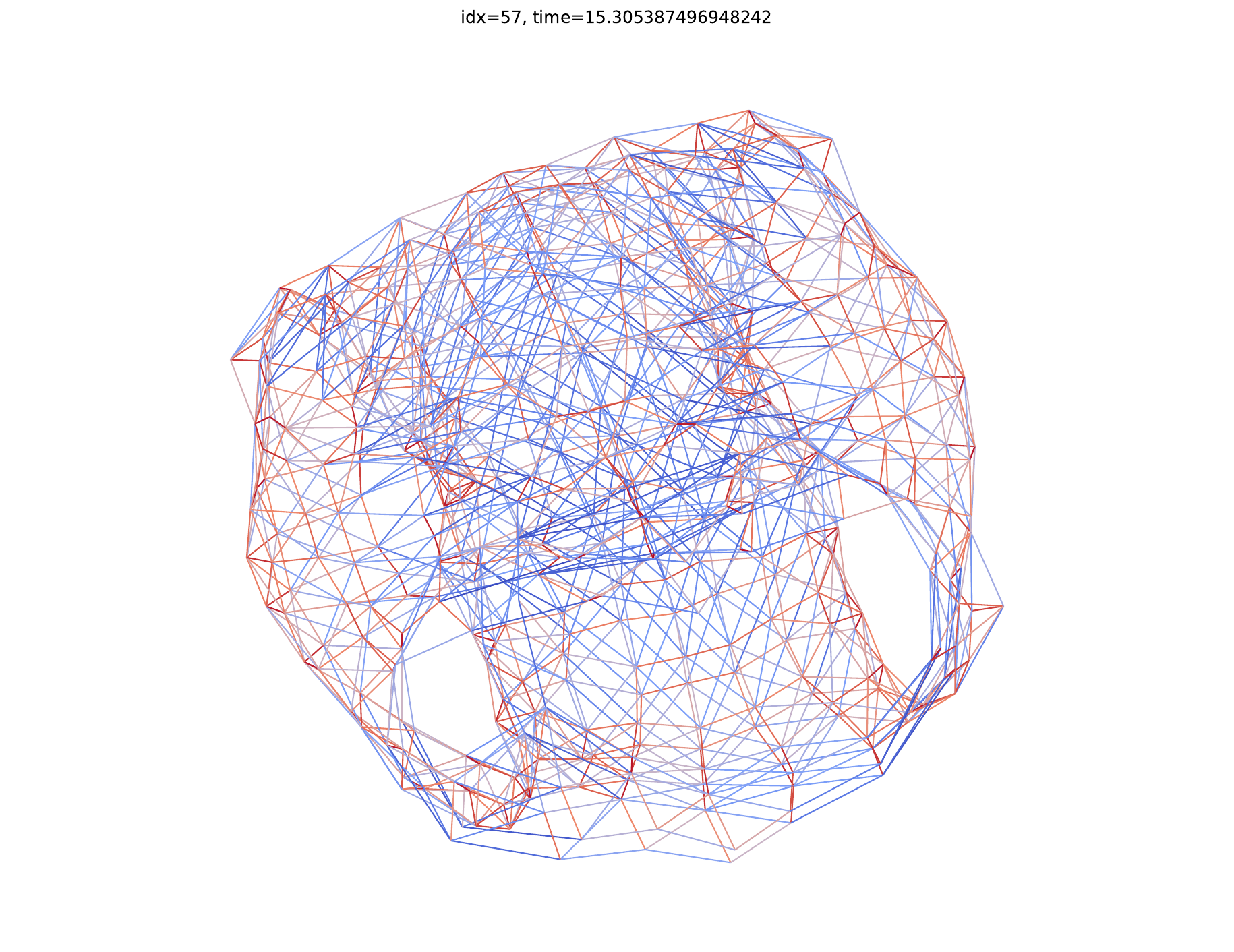} &
\imgcell{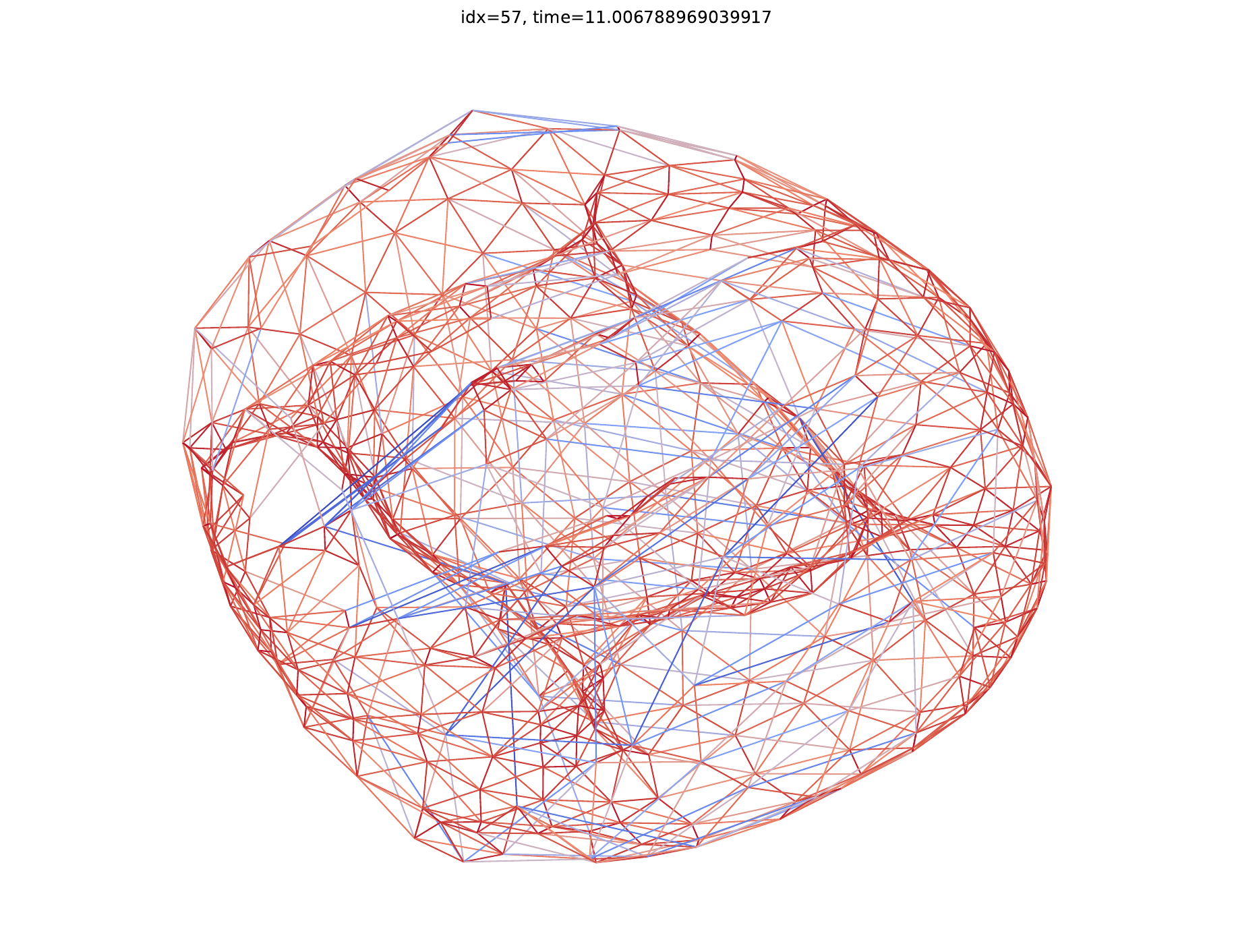} &
\imgcell{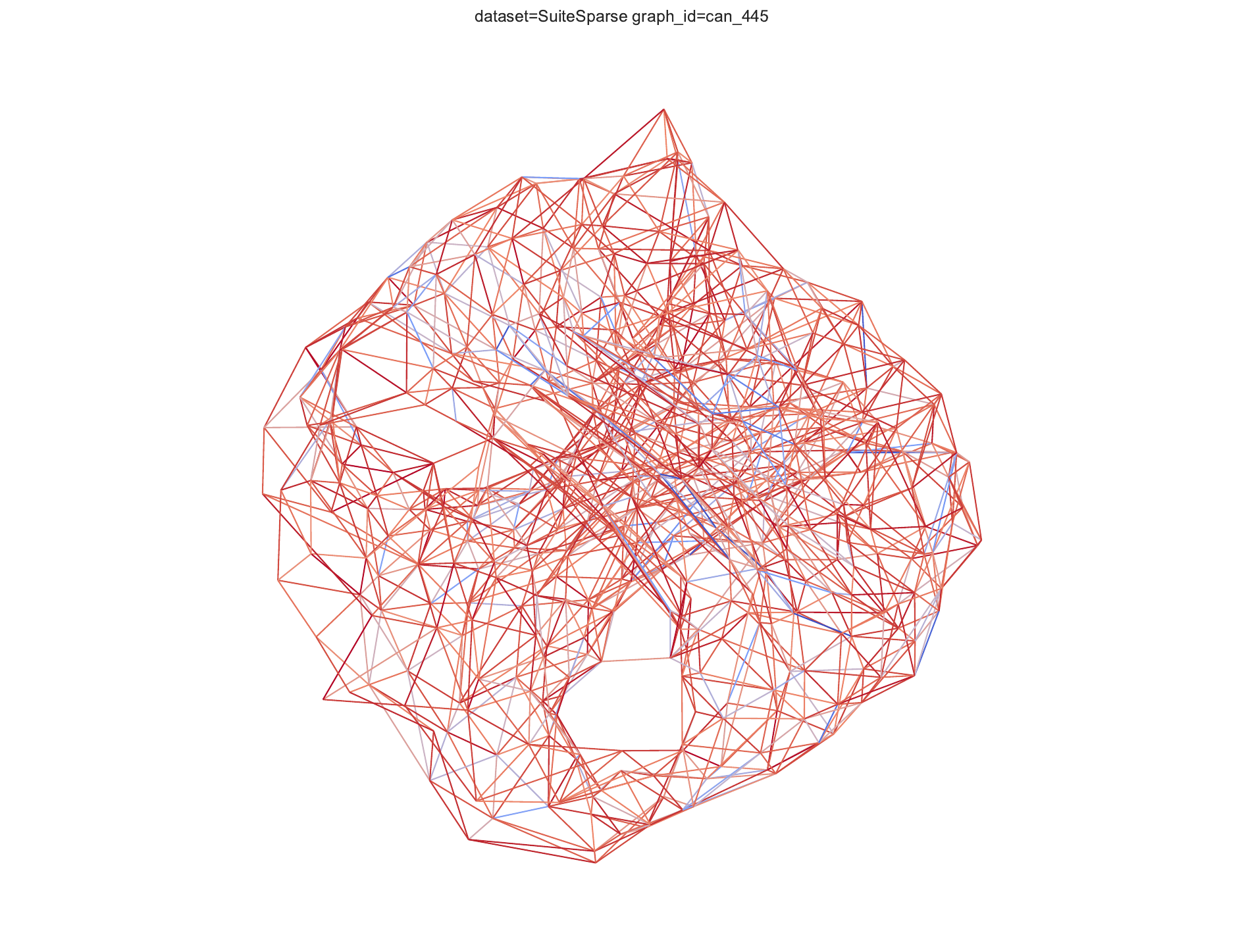} &
\imgcell{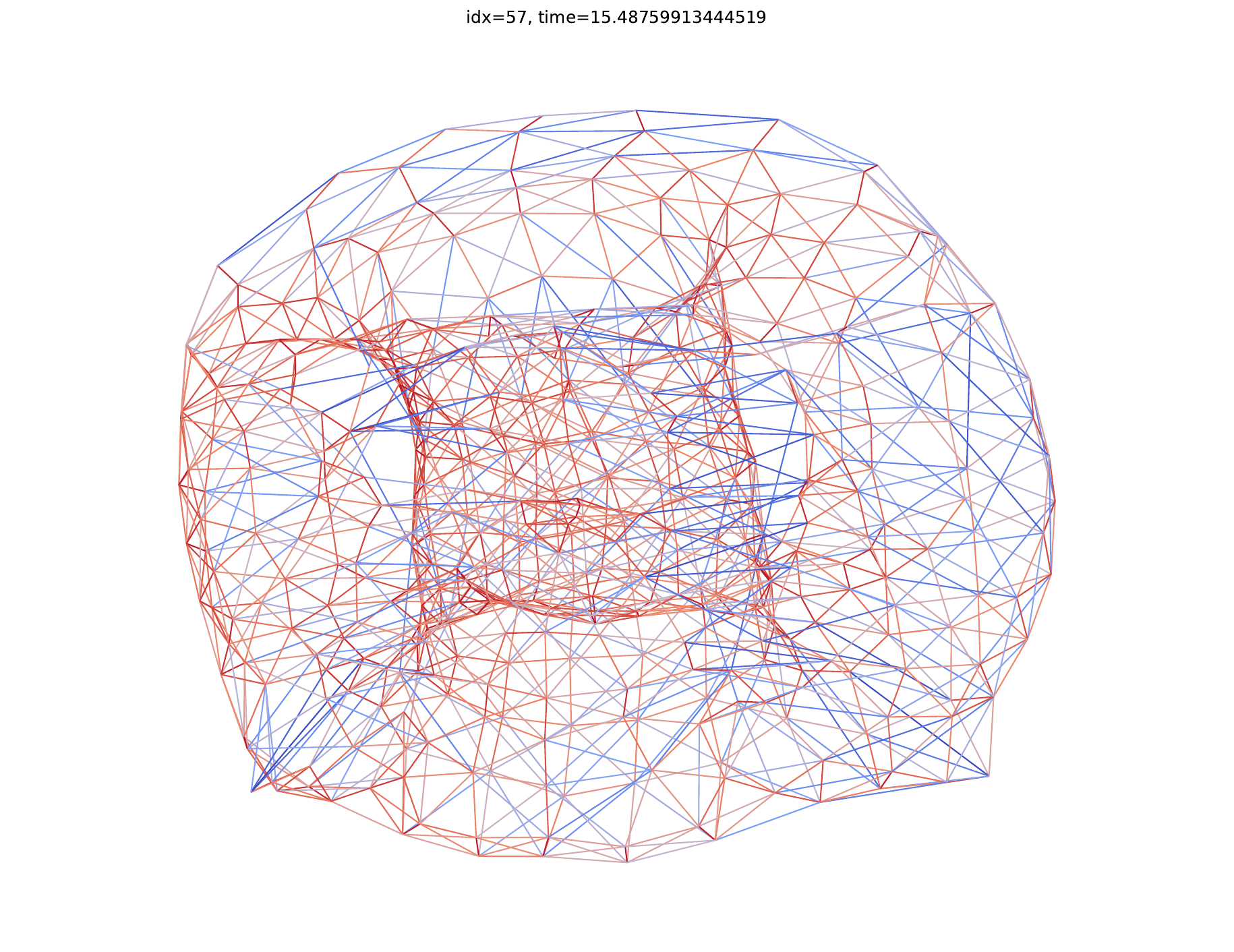} &
\imgcell{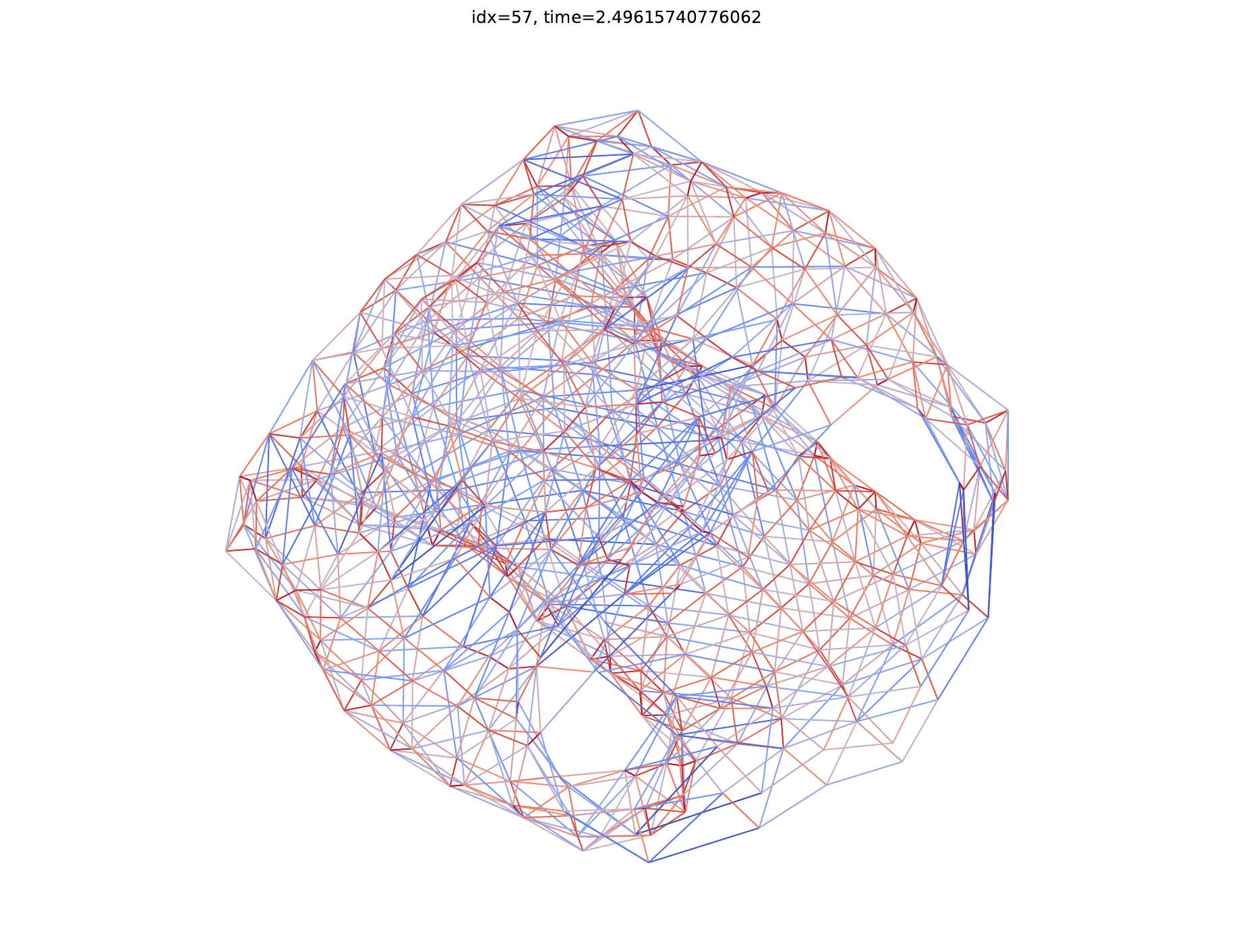} &
\imgcell{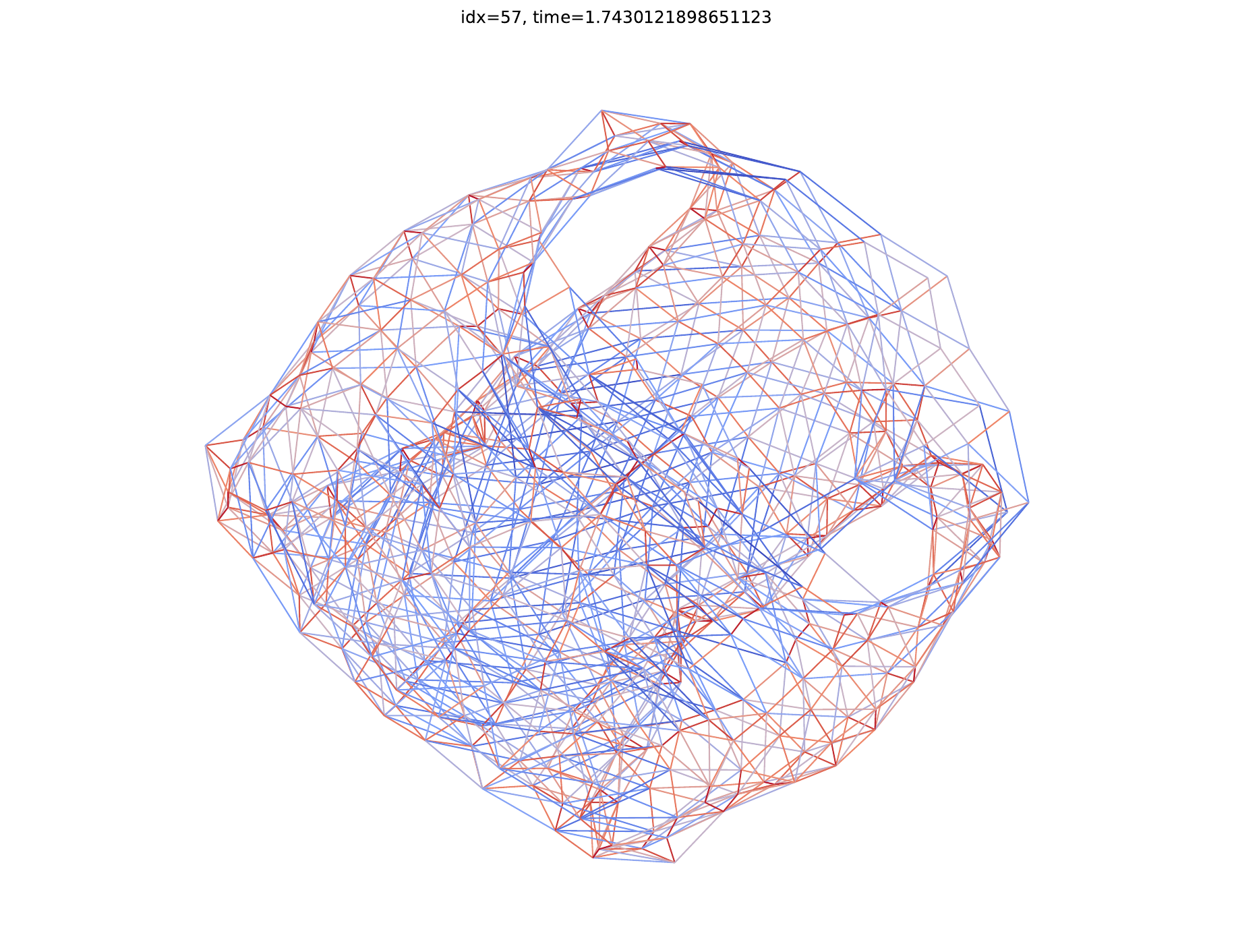} &
\imgcell{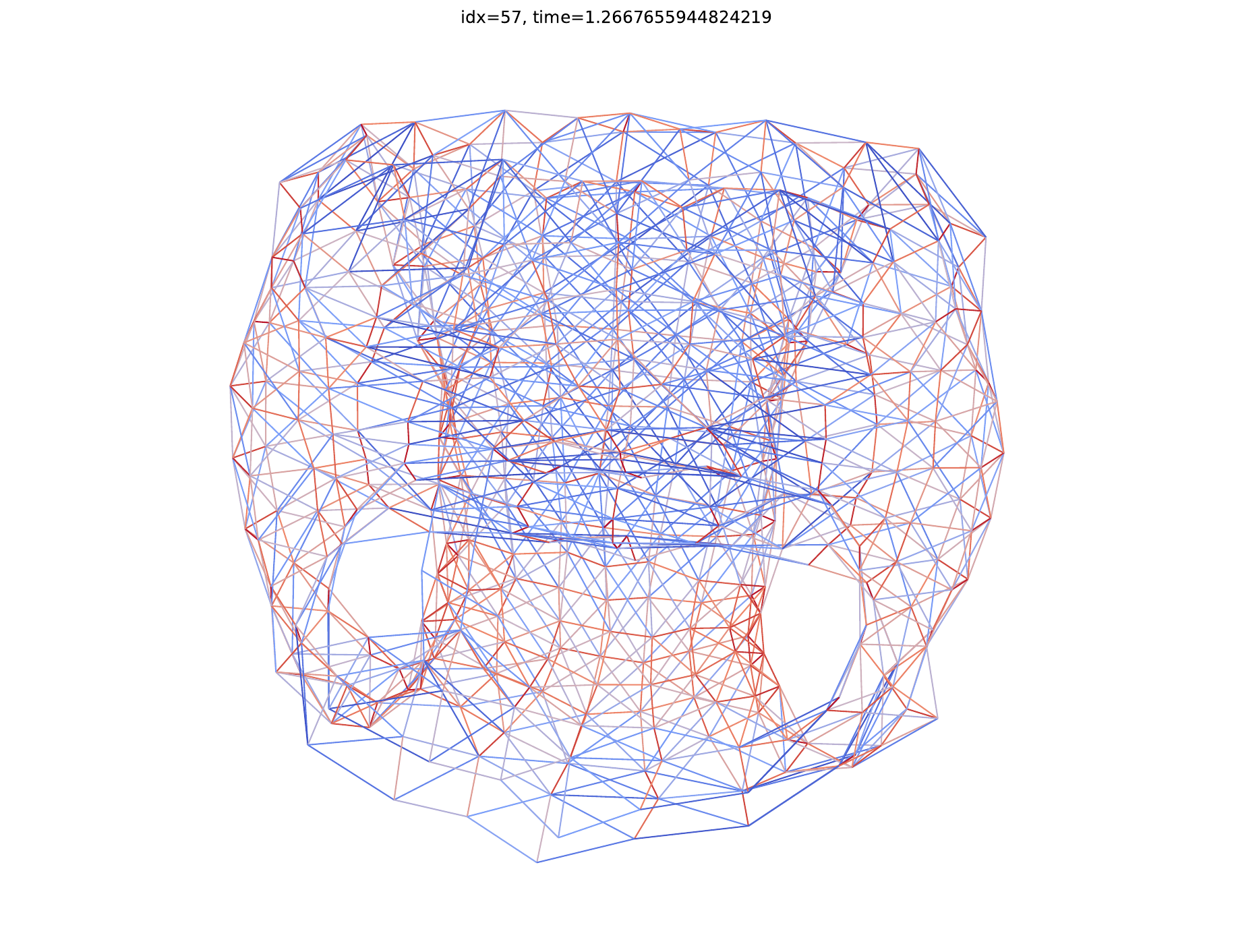} \\

&
t = 0.07s &
t = 7.87s &
t = 14.88s &
t = 0.88s &
t = 7200.00s &
t = 0.67s &
t = 0.77s &
t = 0.82s &
t = 0.76s &
t = 0.80s &
t = 0.66s &
t = 0.85s \\

\makecell{\bfseries nos5\\N = 468\\M = 2352} &
\imgcell{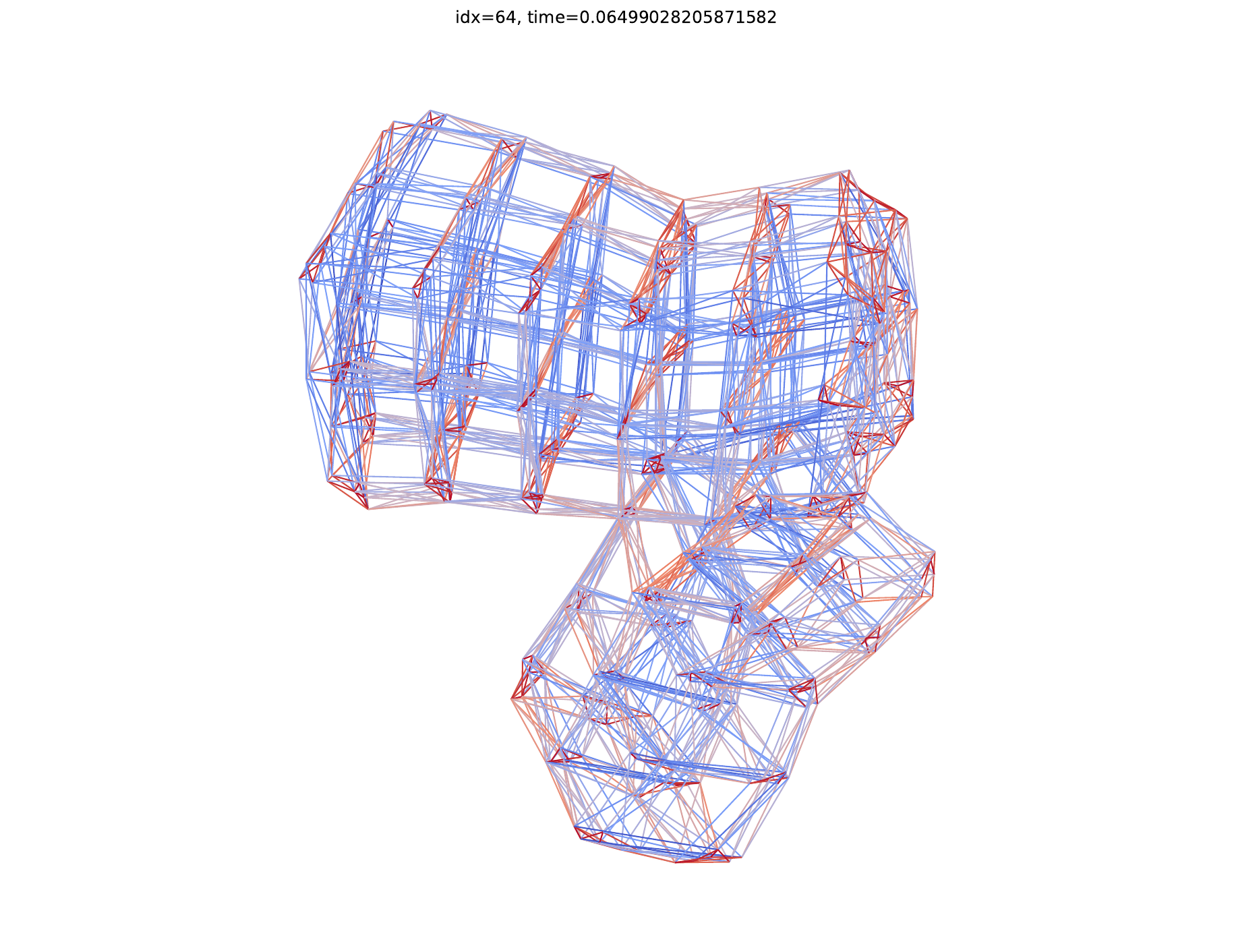} &
\imgcell{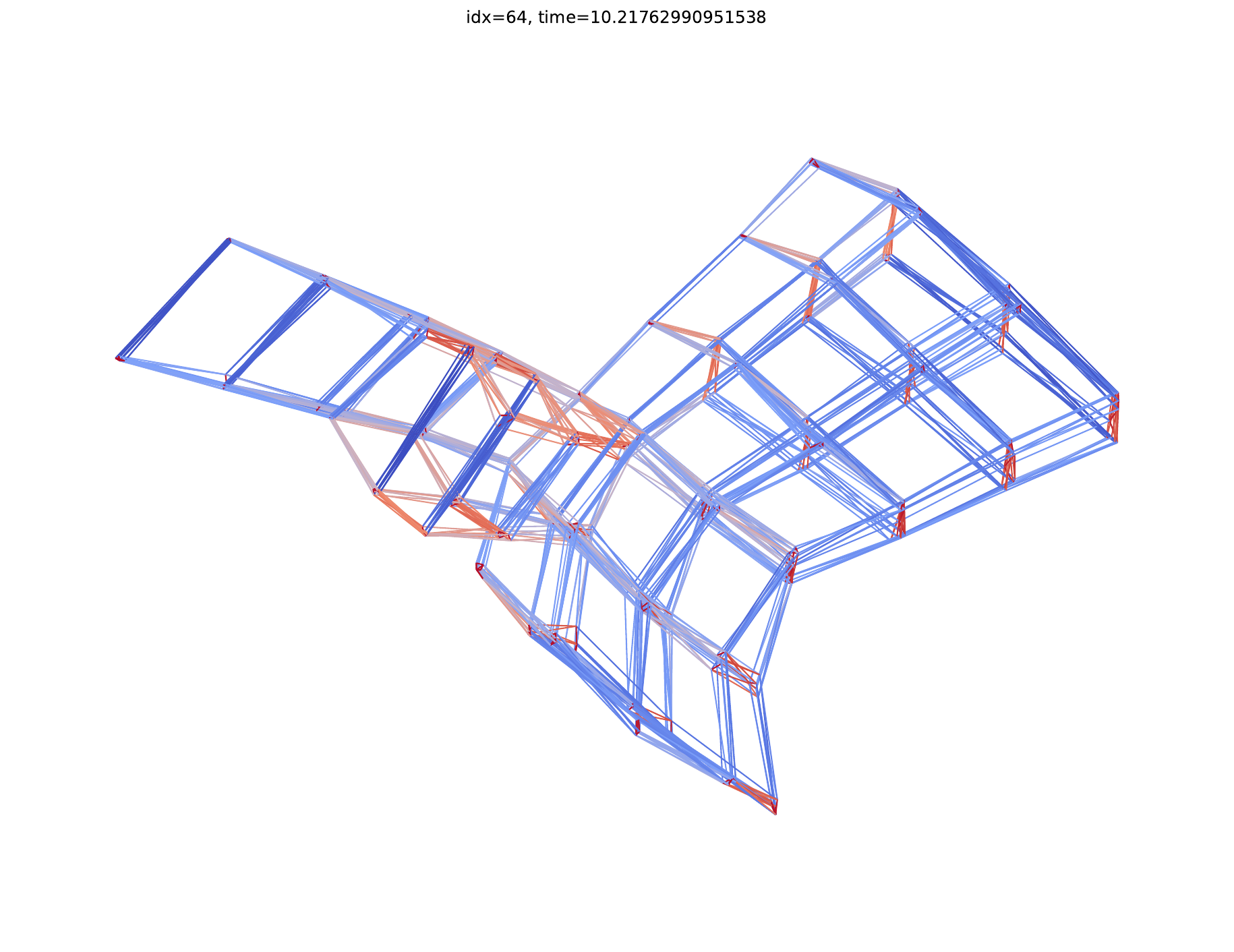} &
\imgcell{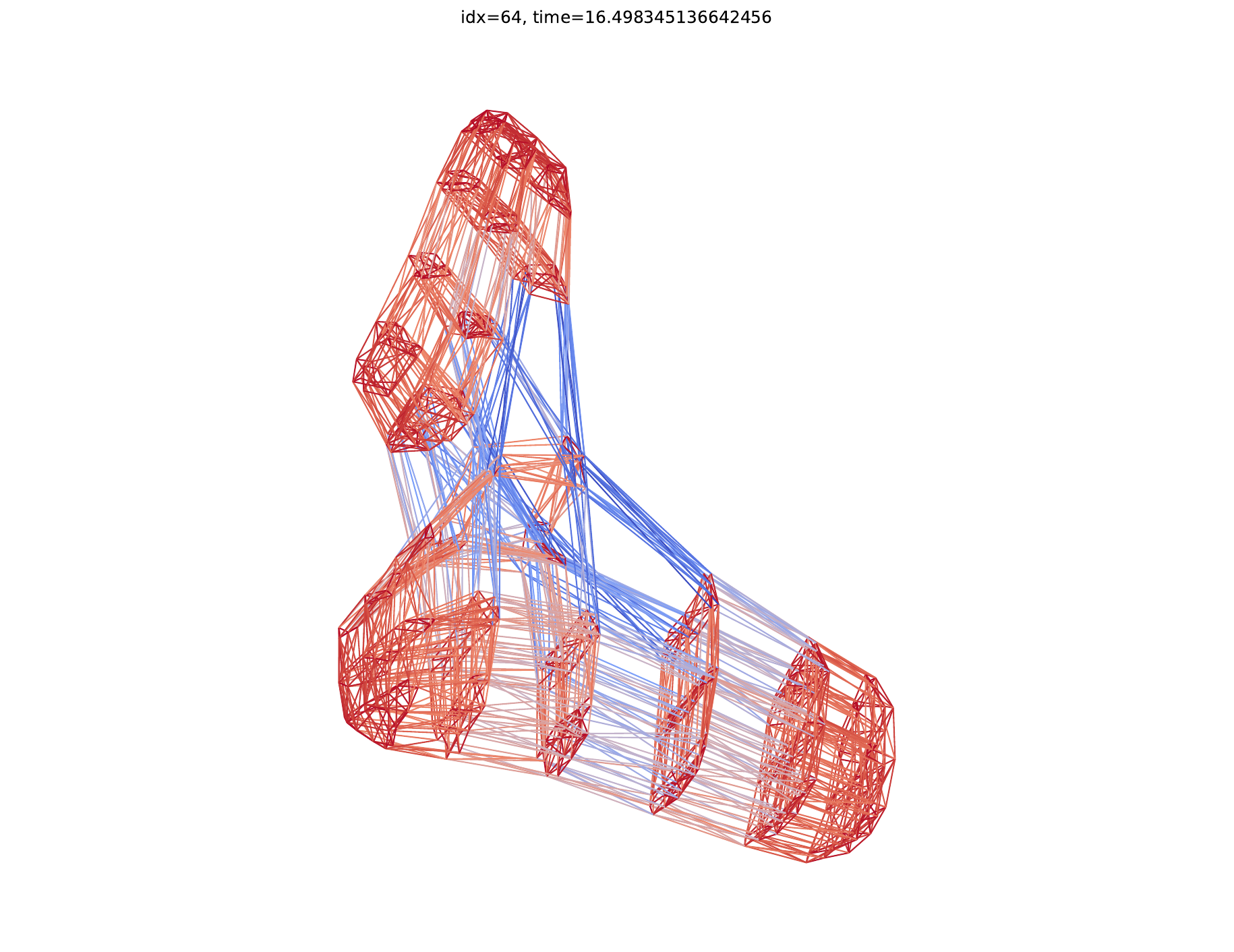} &
\imgcell{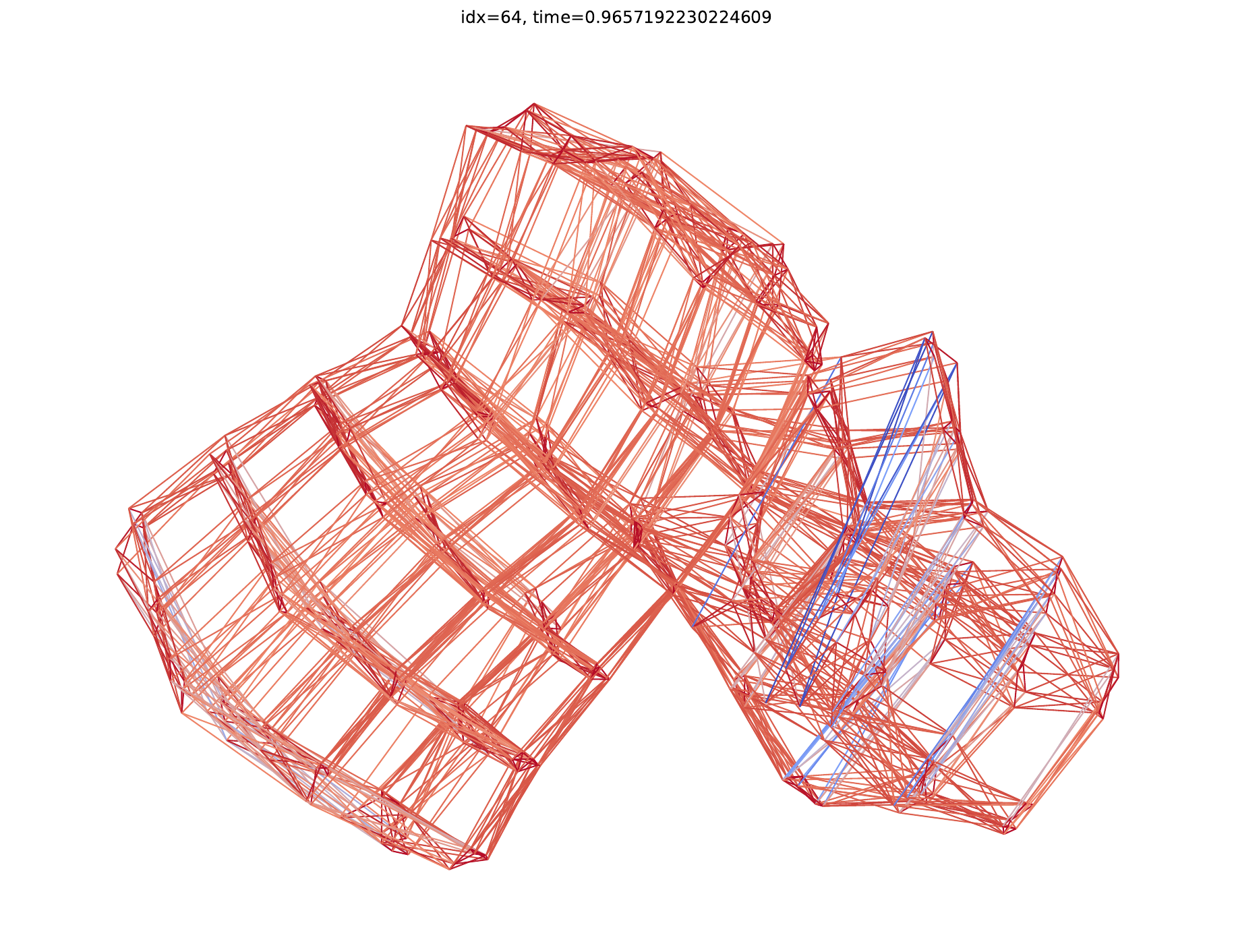} &
\imgcell{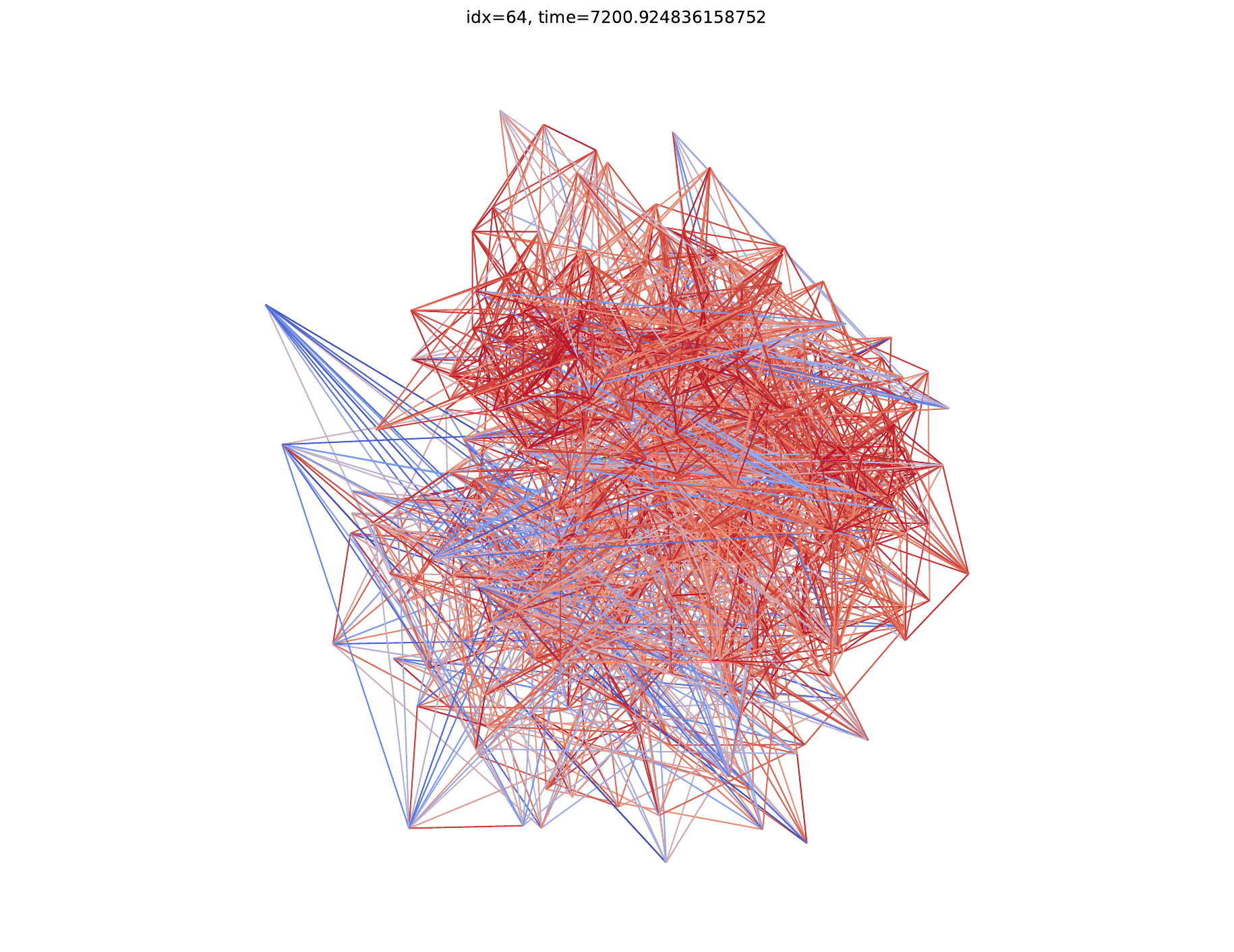} &
\imgcell{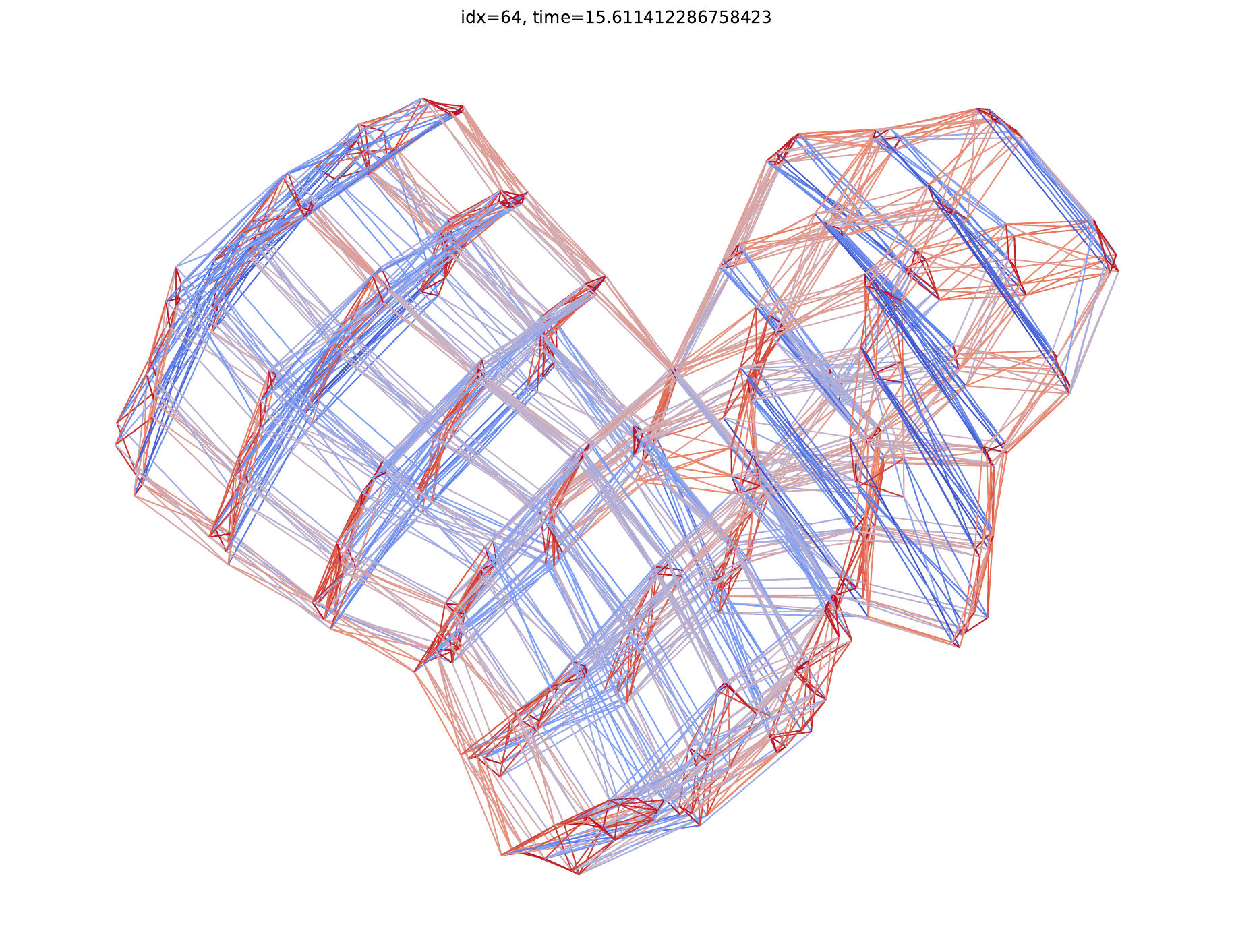} &
\imgcell{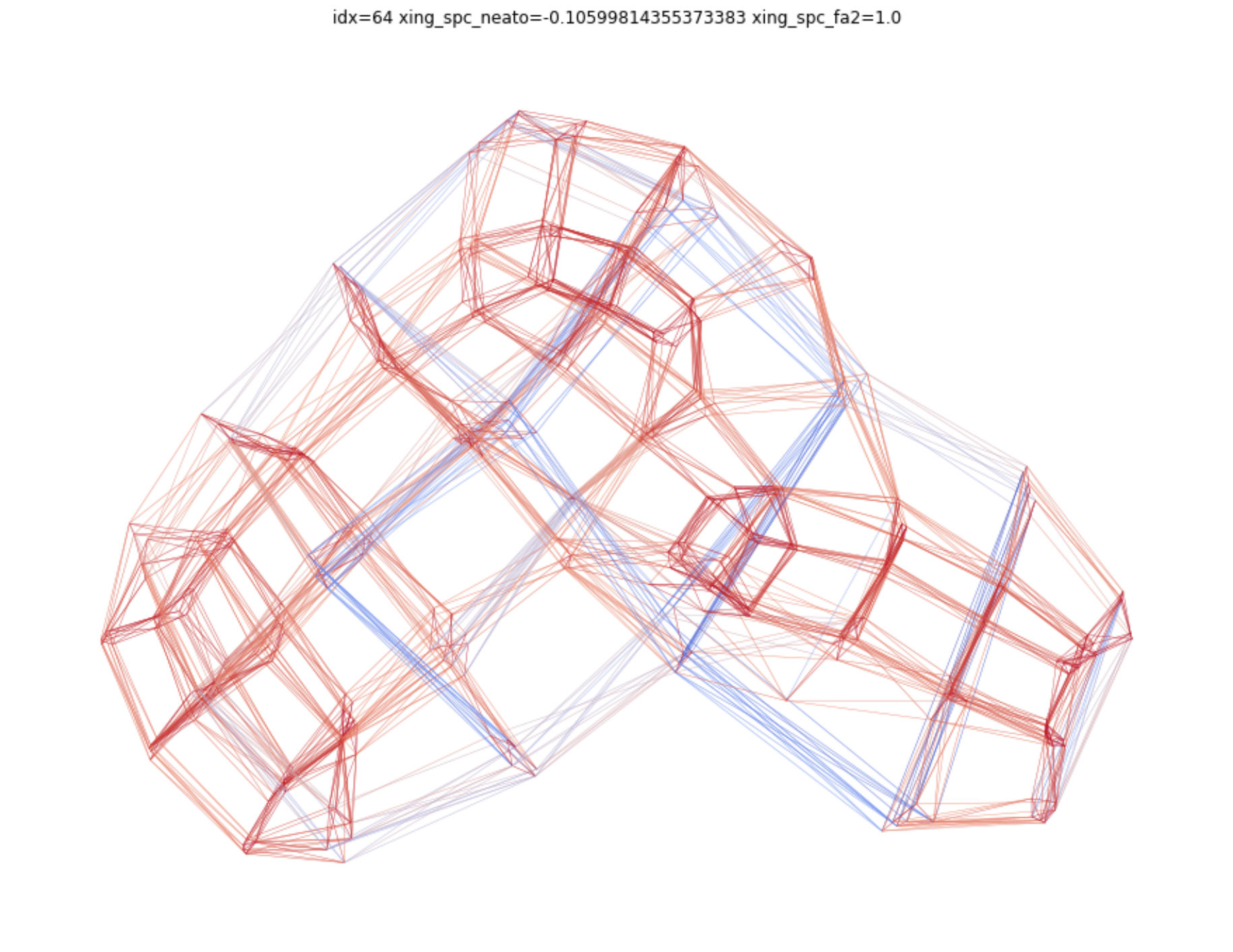} &
\imgcell{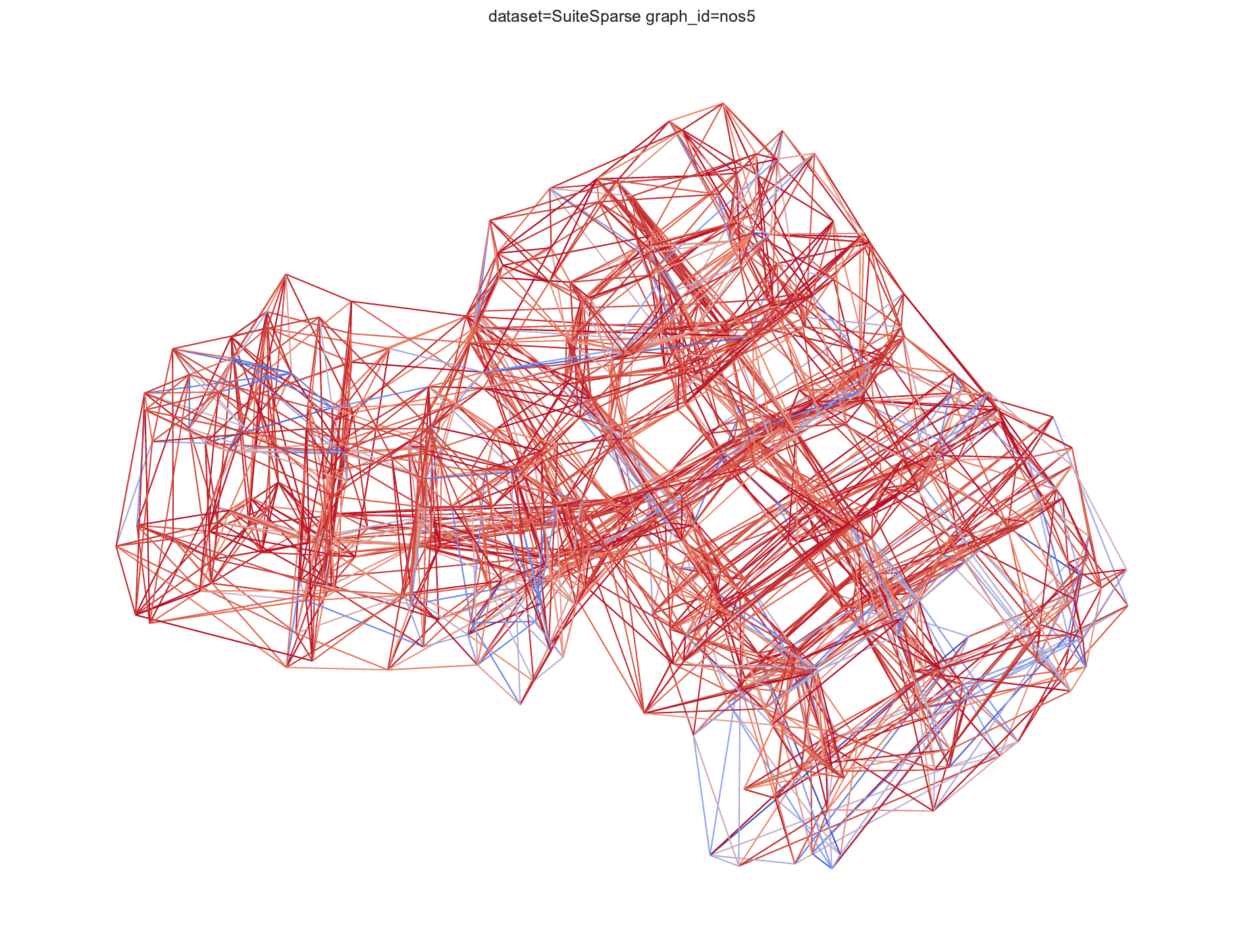} &
\imgcell{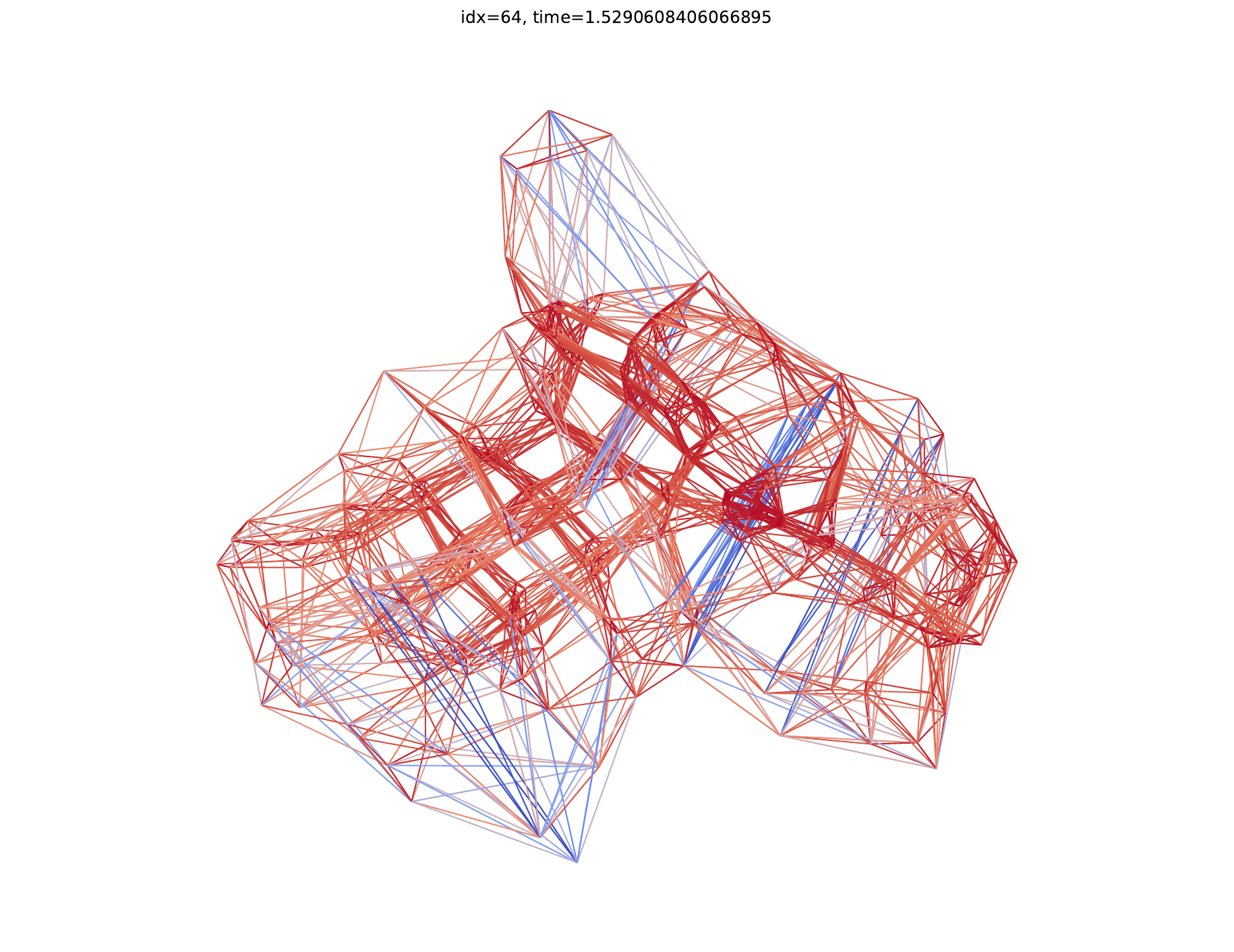} &
\imgcell{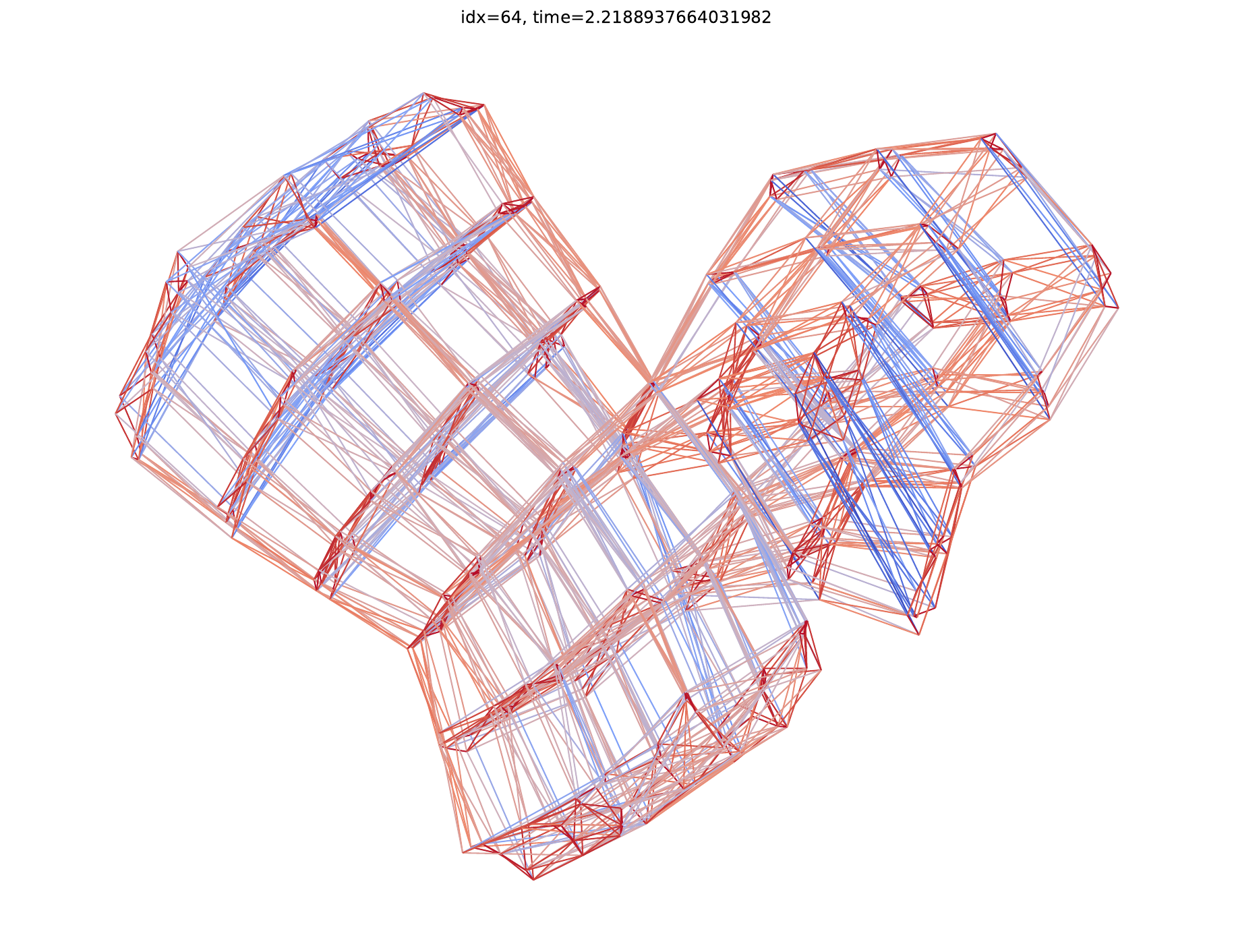} &
\imgcell{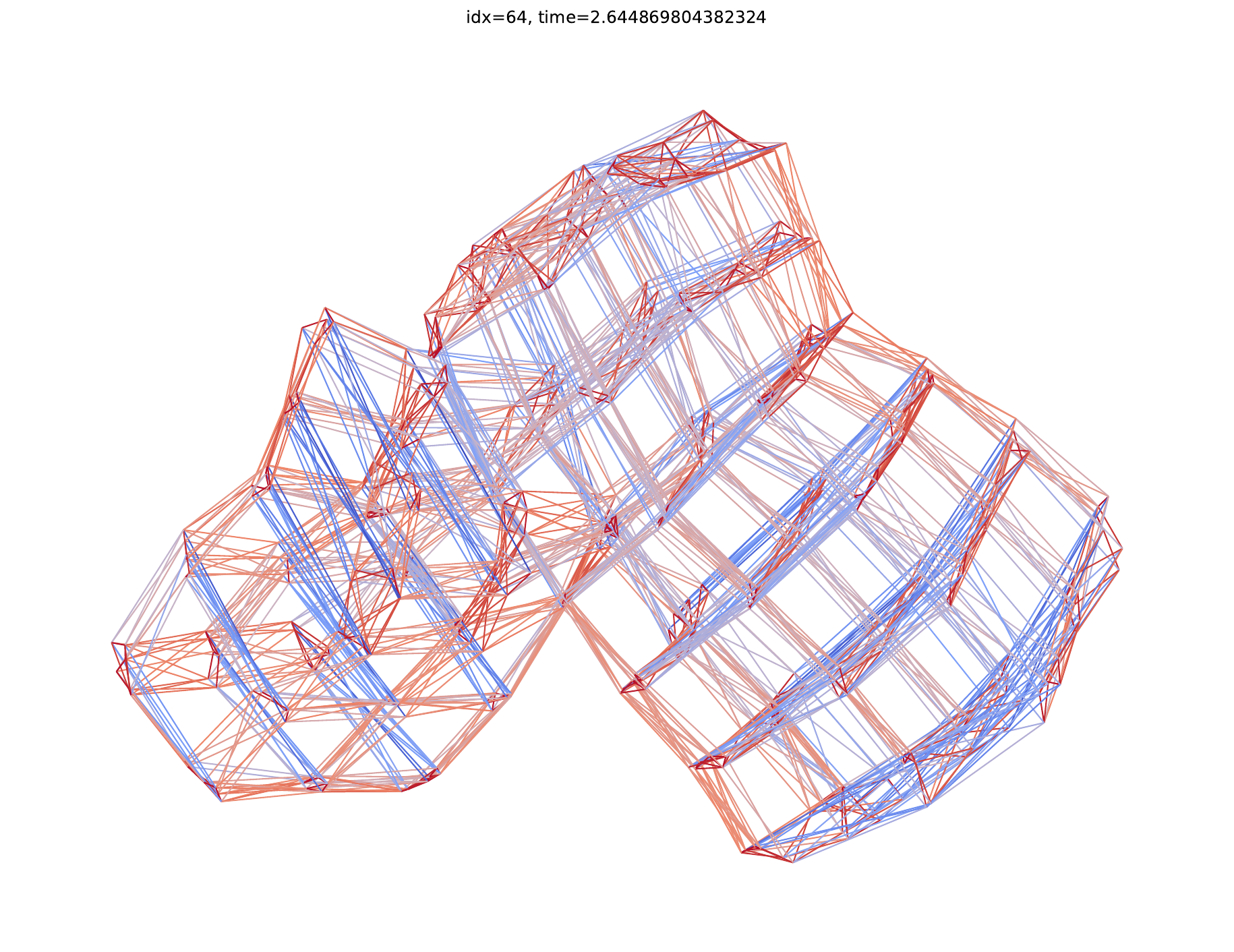} &
\imgcell{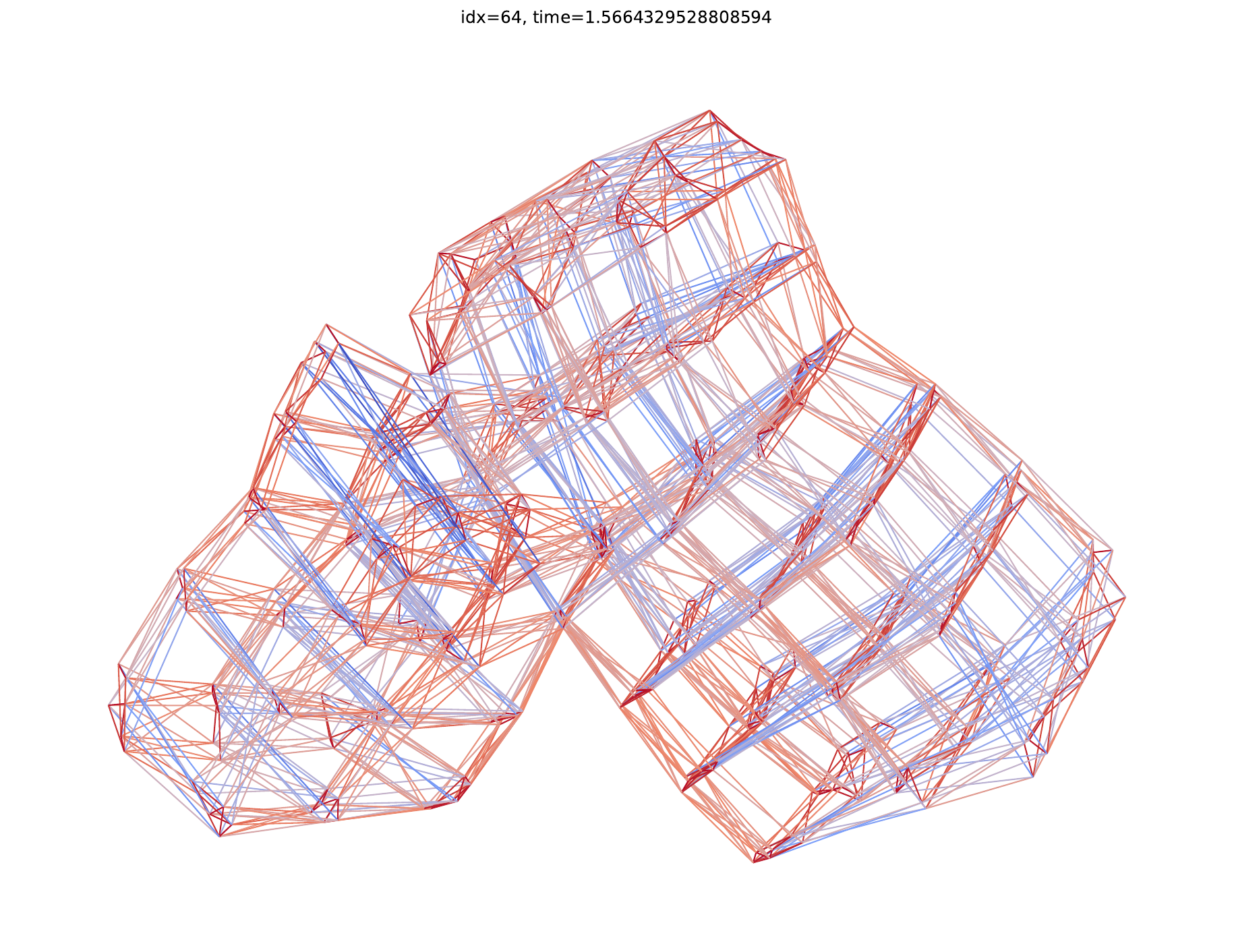} \\

&
t = 0.06s &
t = 10.22s &
t = 16.50s &
t = 0.97s &
t = 7200.00s &
t = 0.86s &
t = 0.79s &
t = 0.80s &
t = 0.73s &
t = 0.68s &
t = 0.93s &
t = 0.67s \\

\makecell{\bfseries dwt\_503\\N = 503\\M = 2762} &
\imgcell{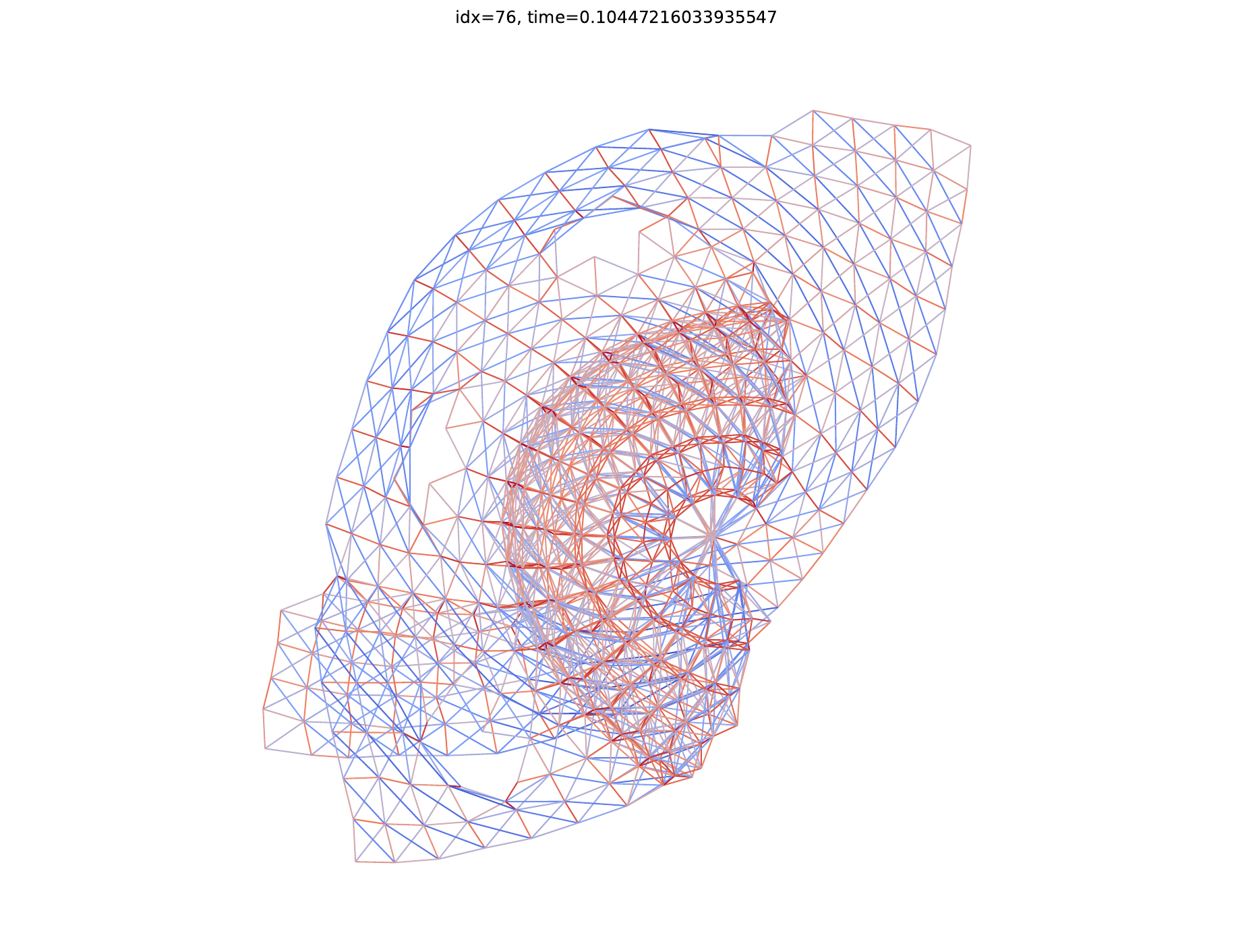} &
\imgcell{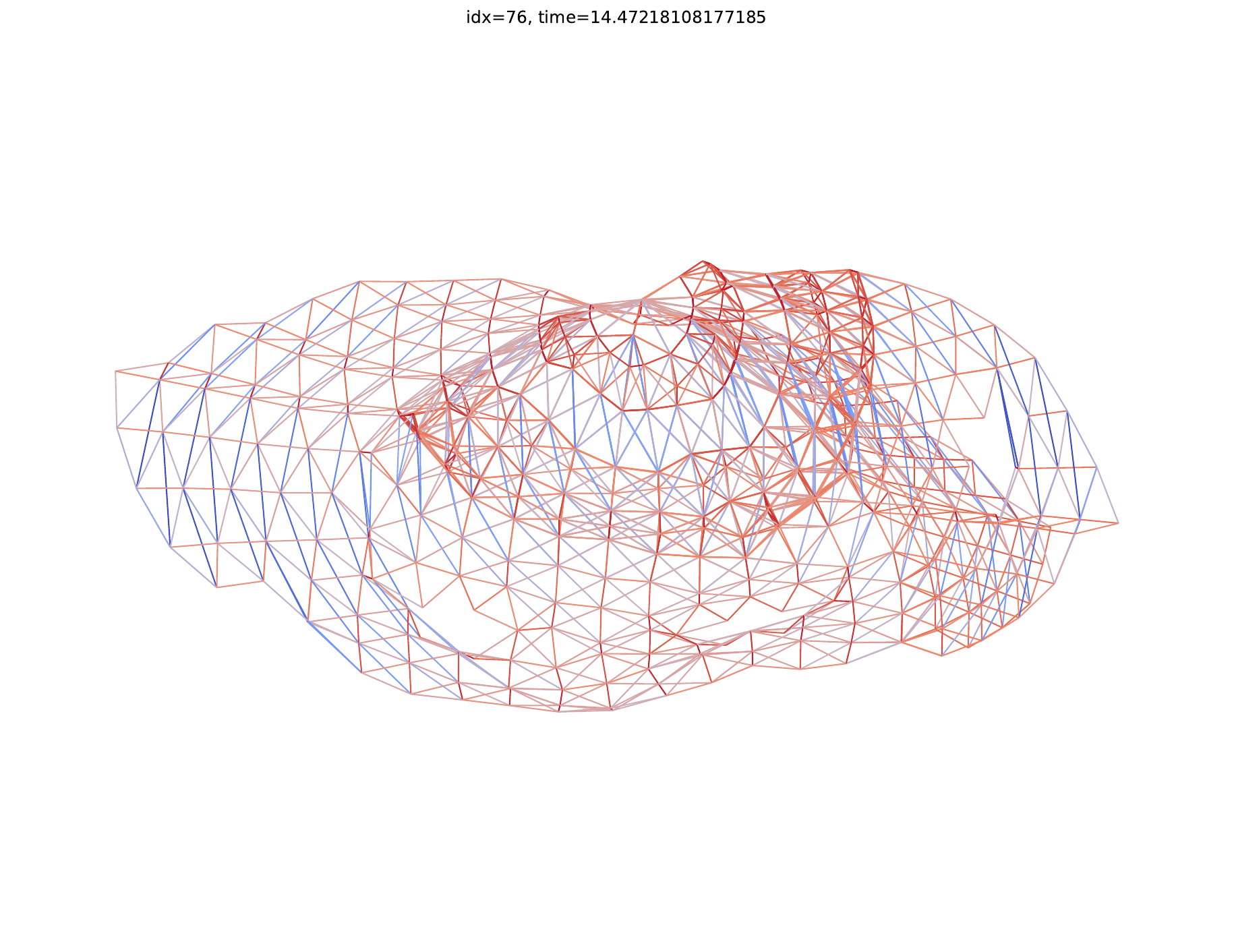} &
\imgcell{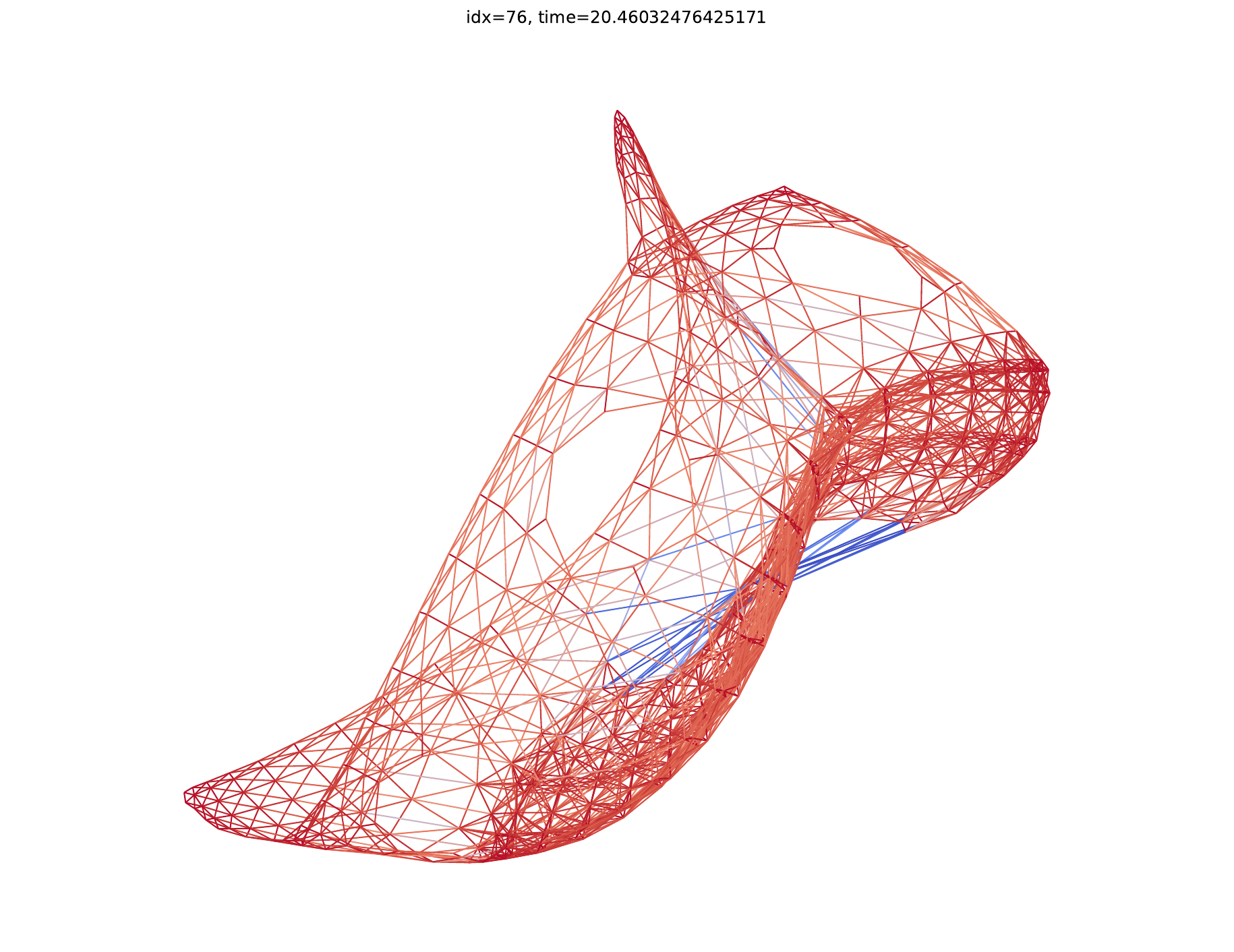} &
\imgcell{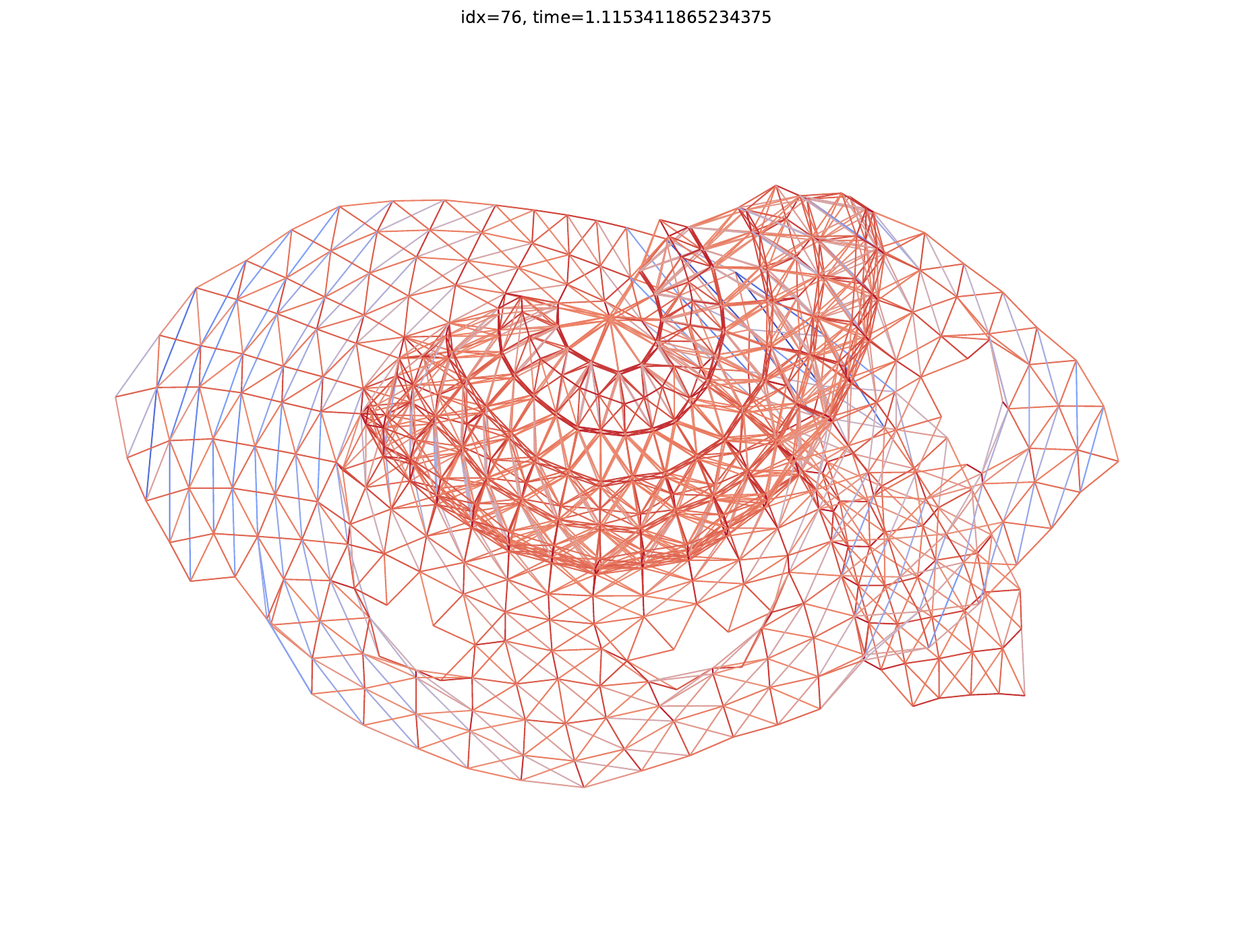} &
\imgcell{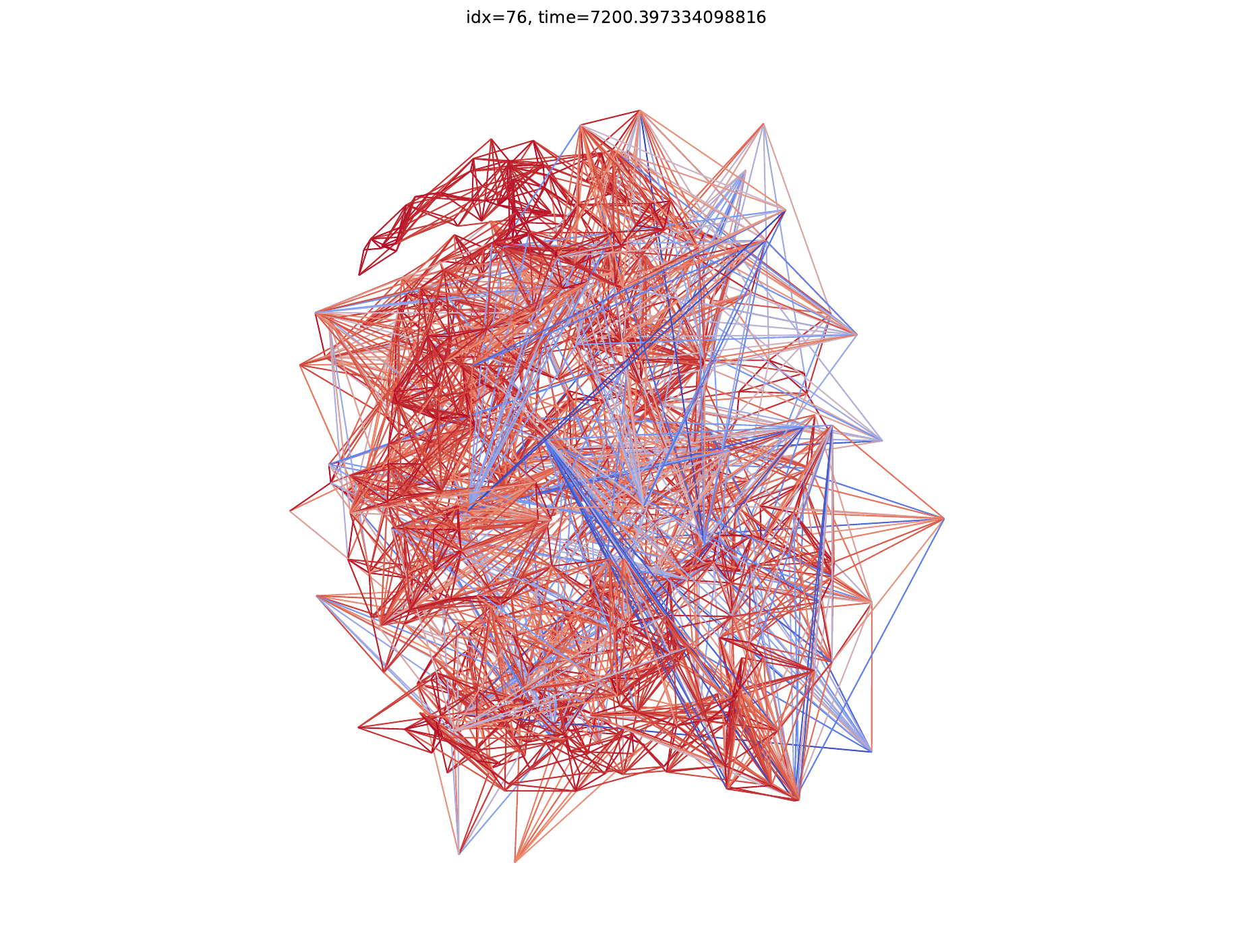} &
\imgcell{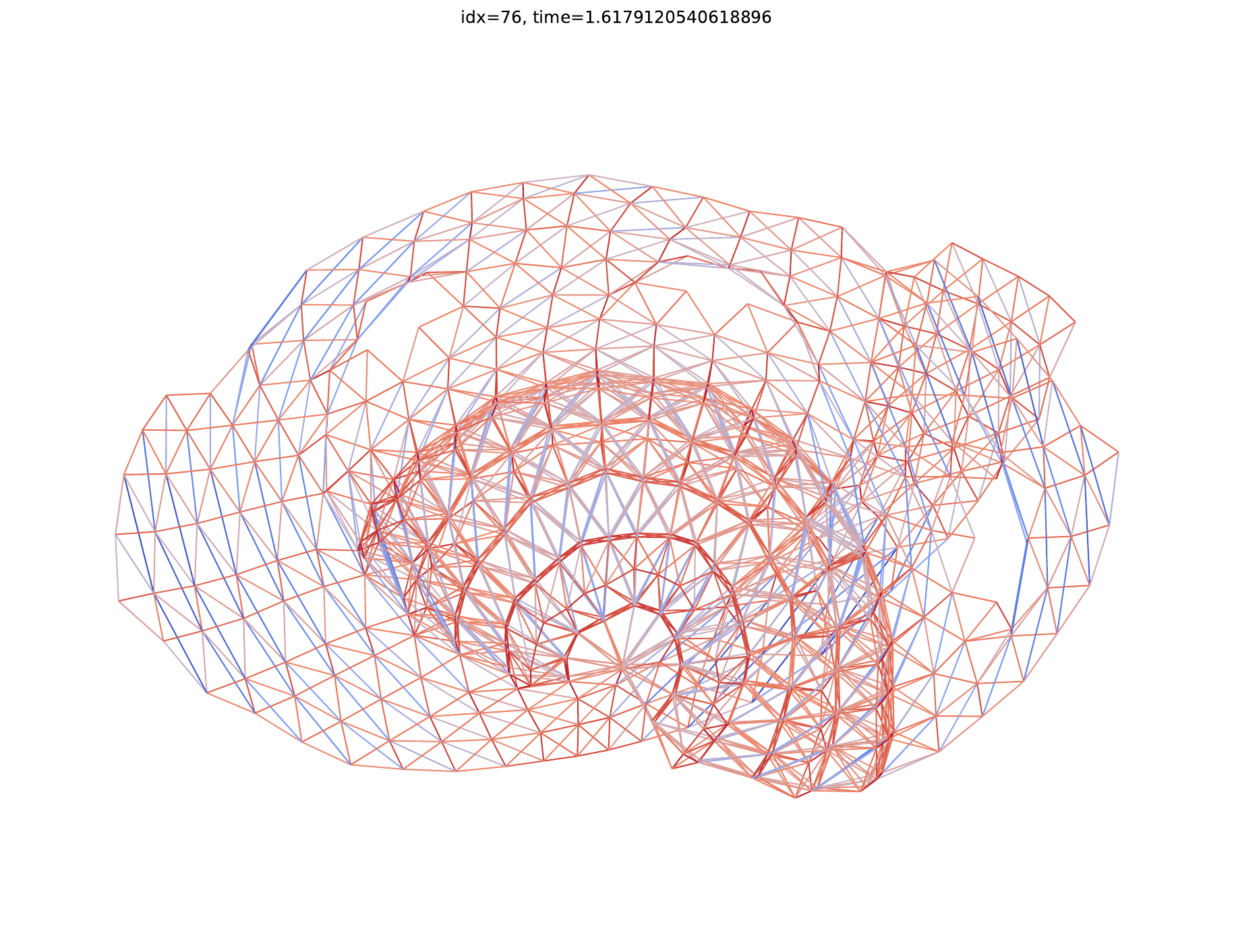} &
\imgcell{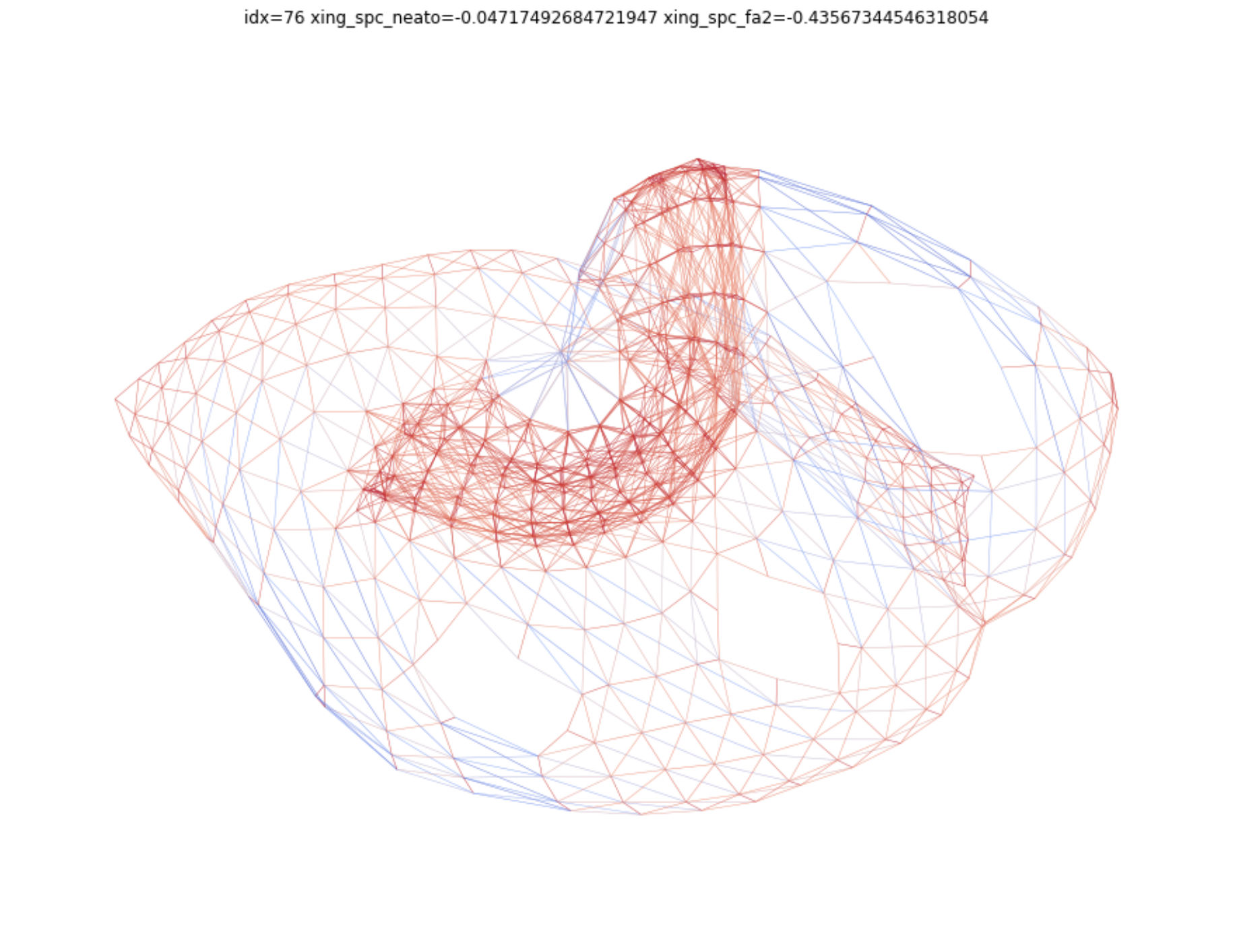} &
\imgcell{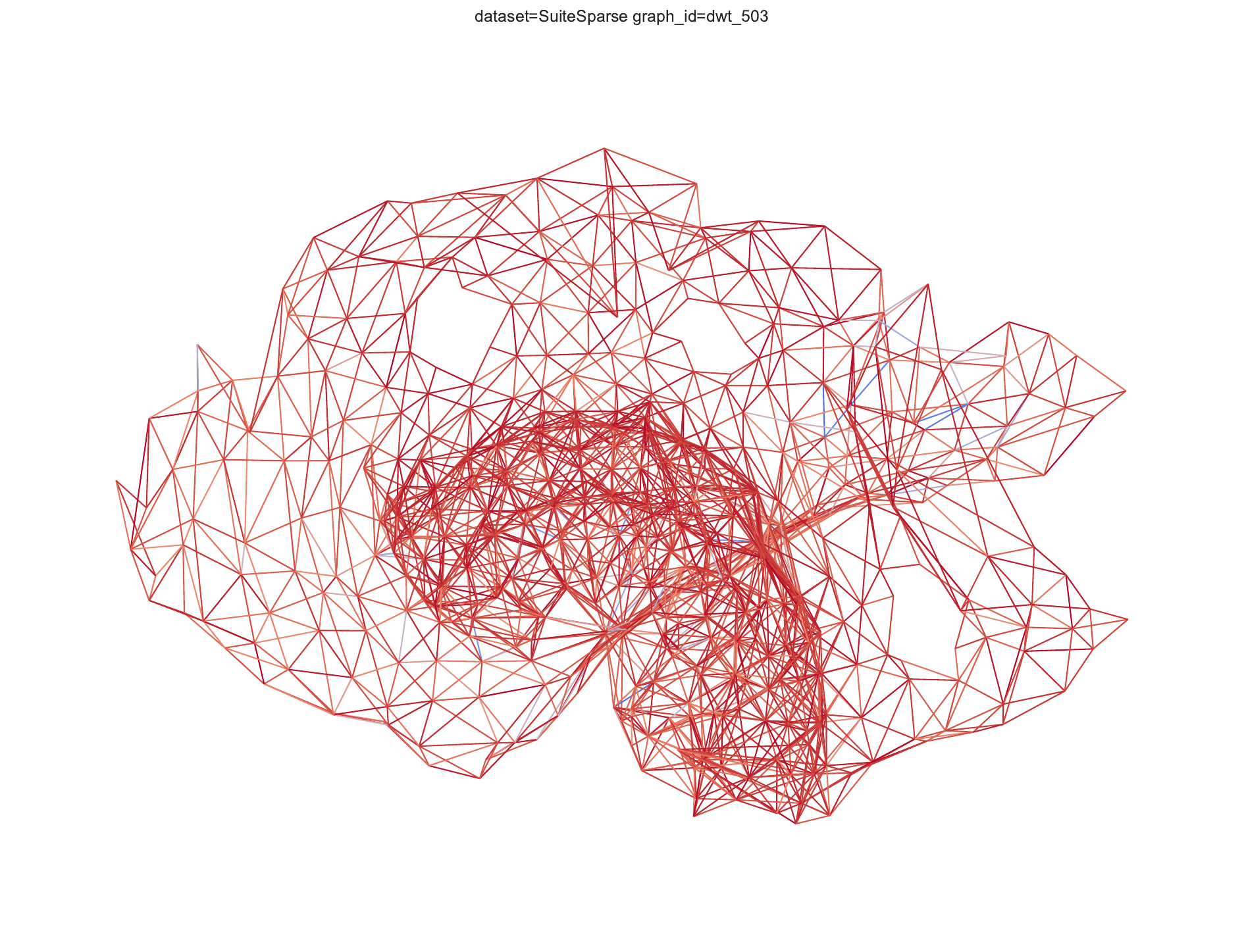} &
\imgcell{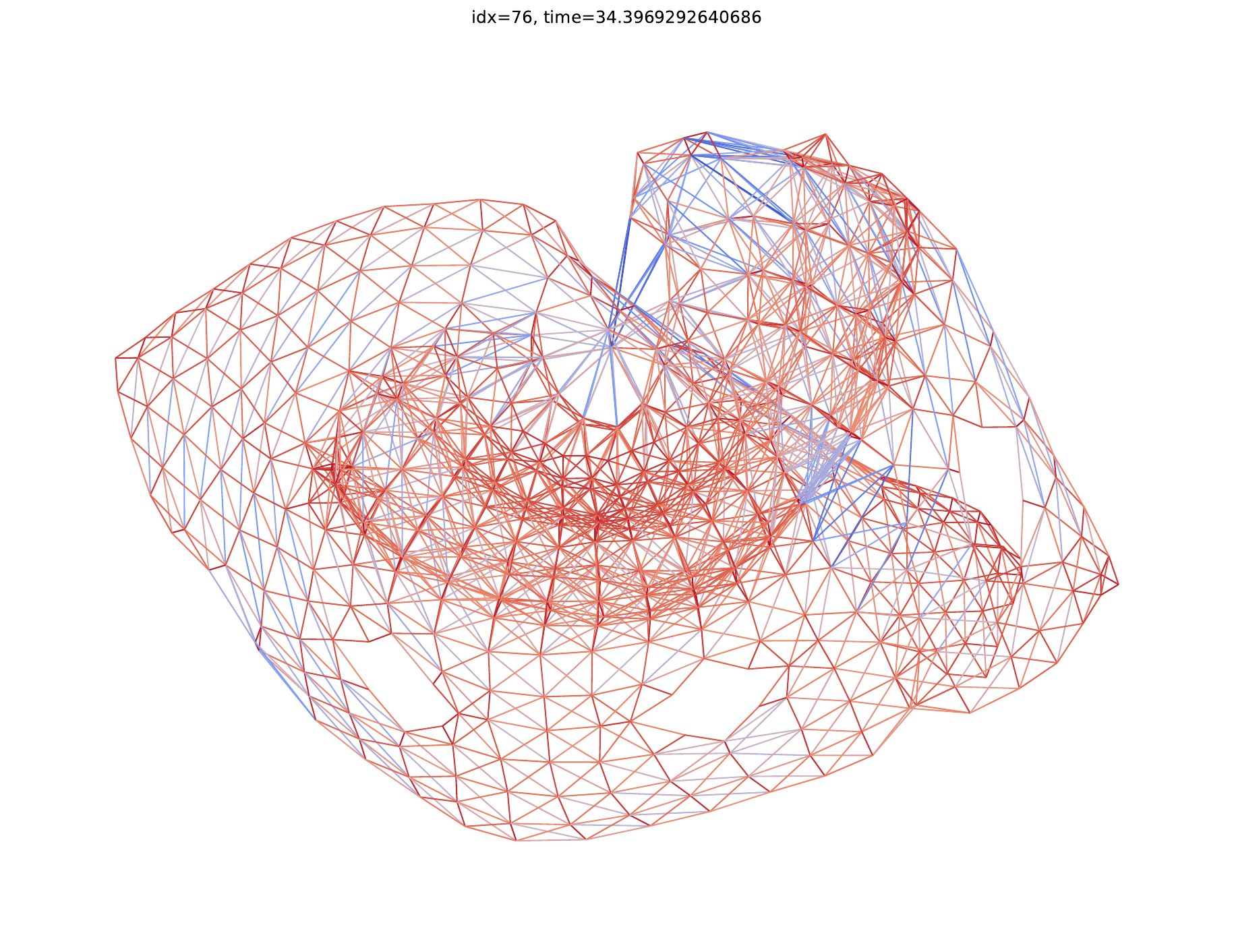} &
\imgcell{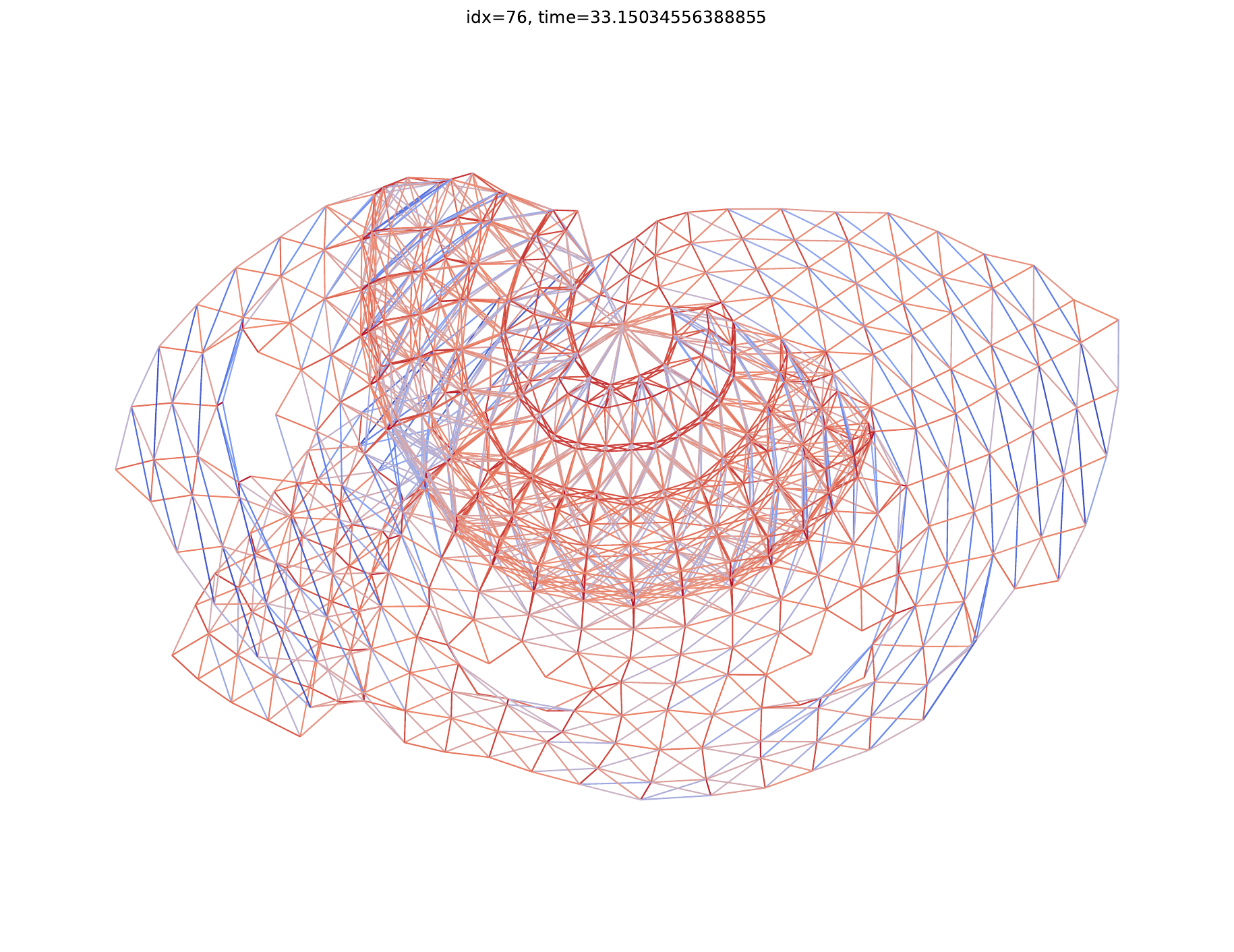} &
\imgcell{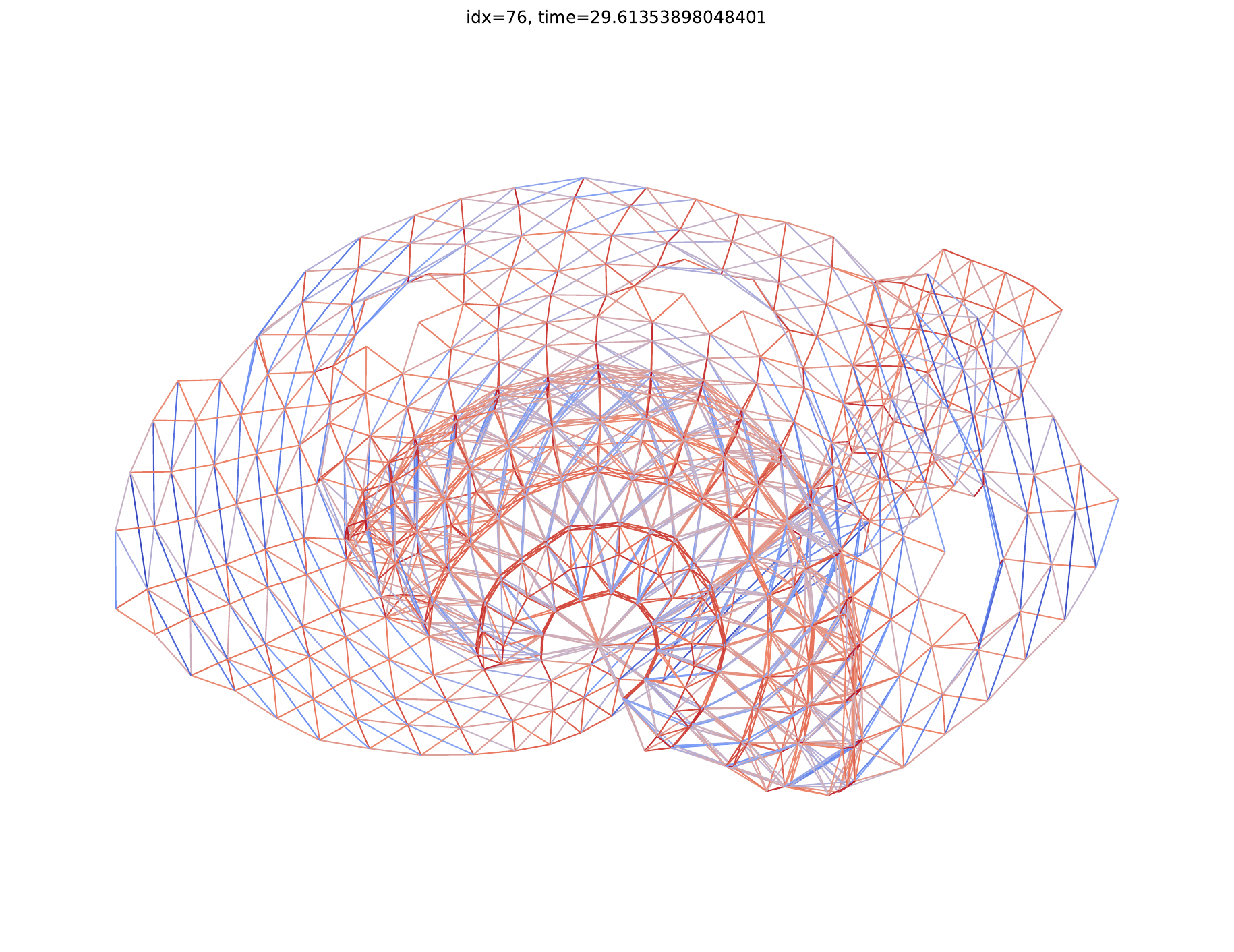} &
\imgcell{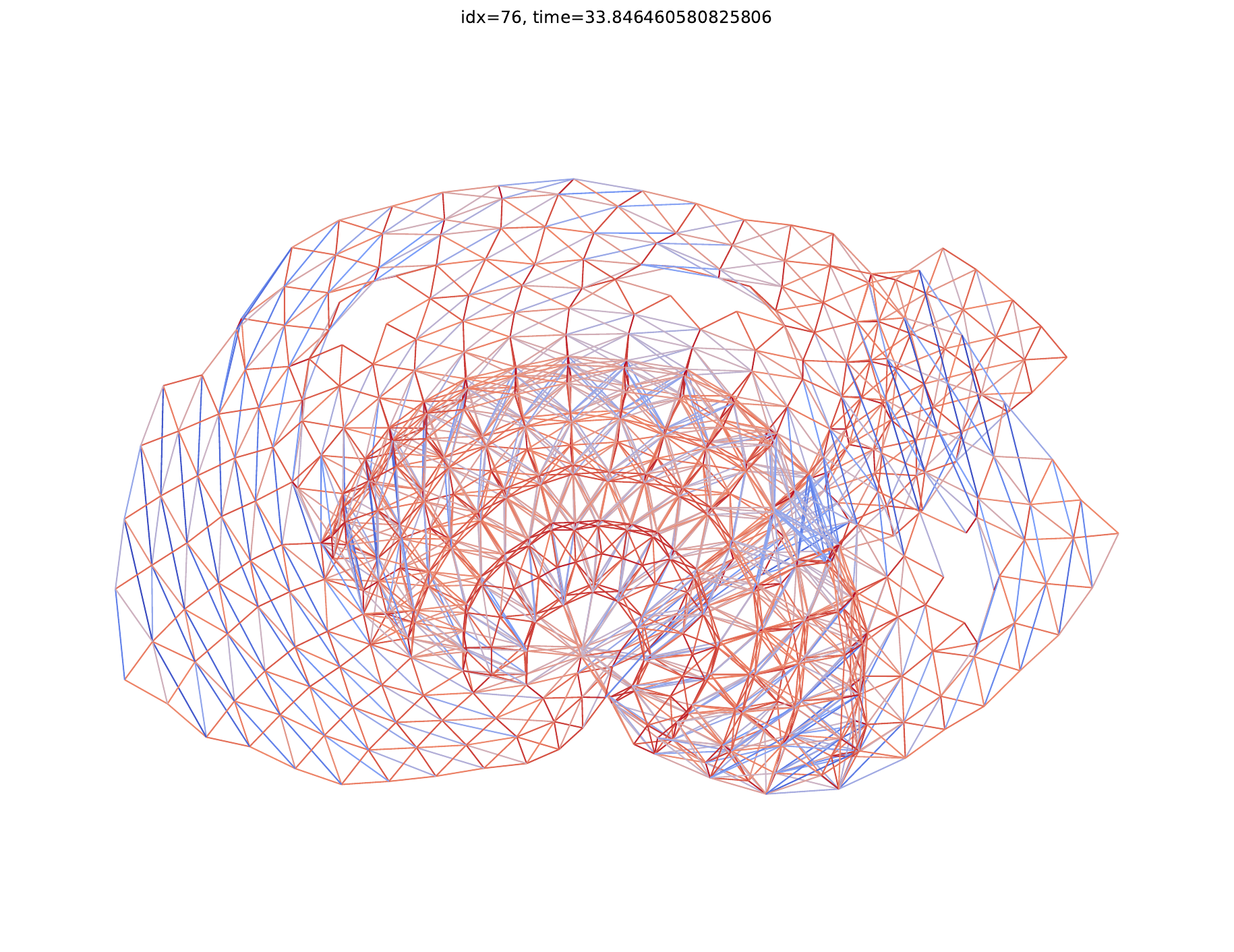} \\

&
t = 0.10s &
t = 14.47s &
t = 20.46s &
t = 1.12s &
t = 7200.00s &
t = 0.92s &
t = 1.20s &
t = 1.07s &
t = 1.00s &
t = 0.85s &
t = 1.02s &
t = 0.89s \\

\makecell{\bfseries Spectro\_10NN\\N = 531\\M = 3711} &
\imgcell{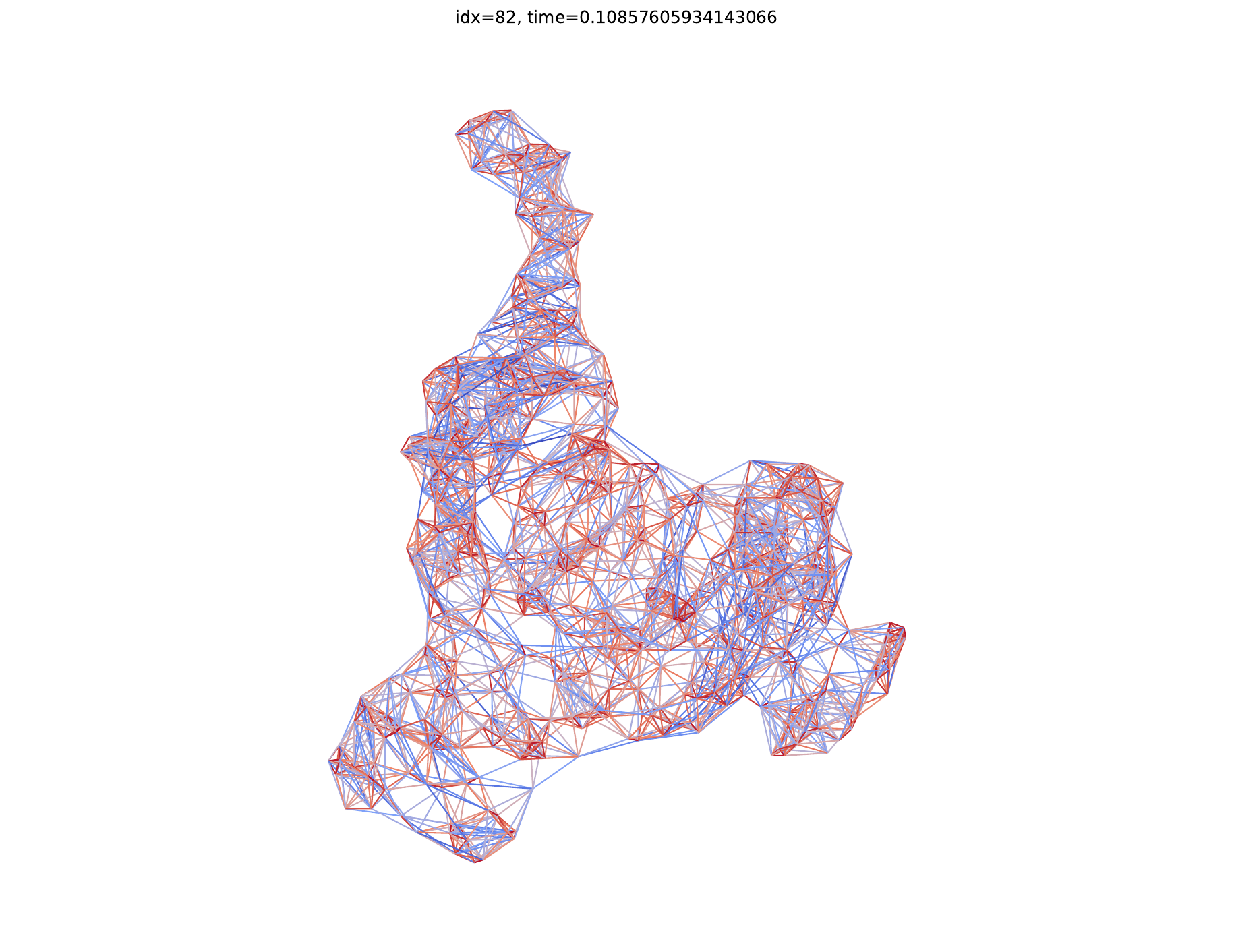} &
\imgcell{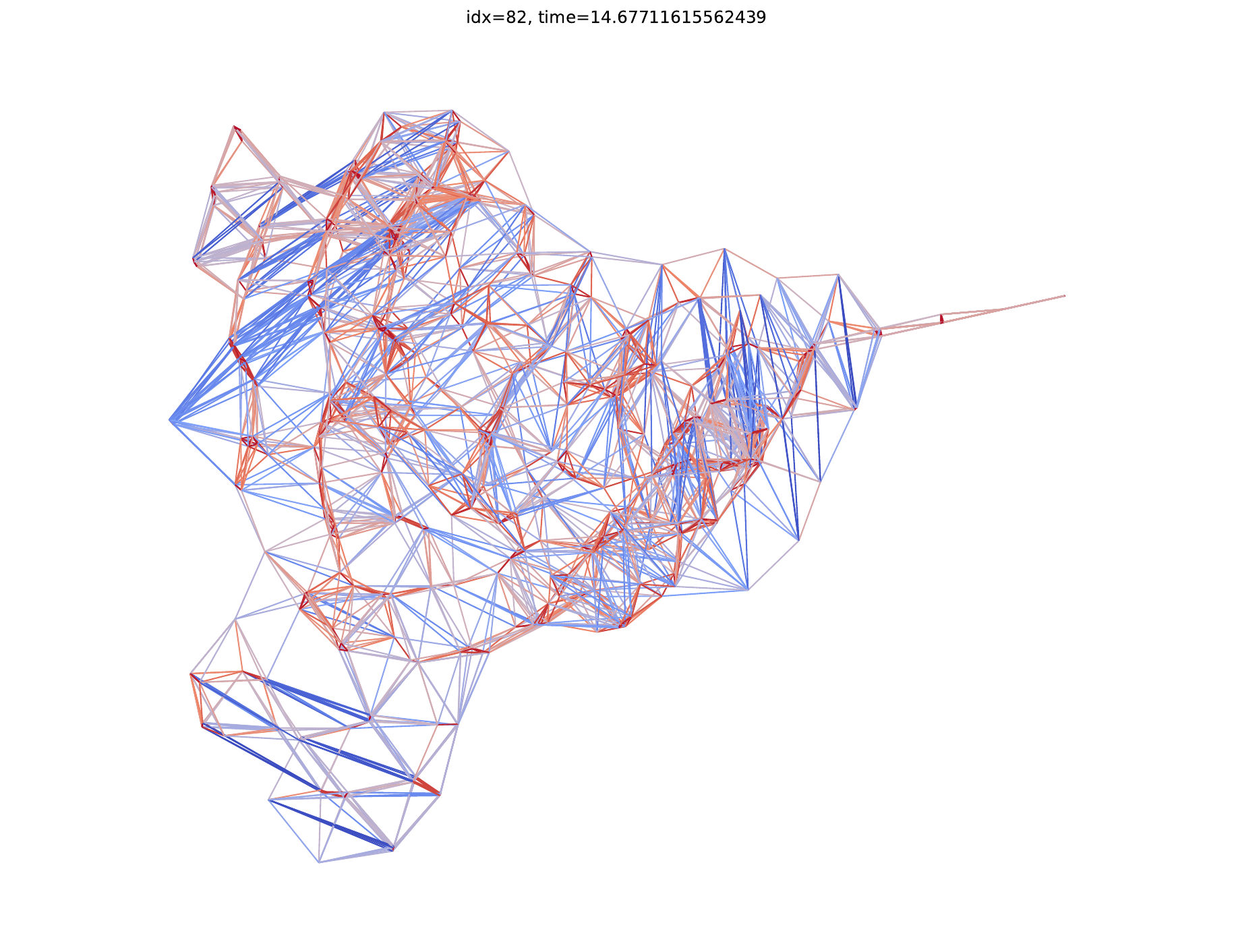} &
\imgcell{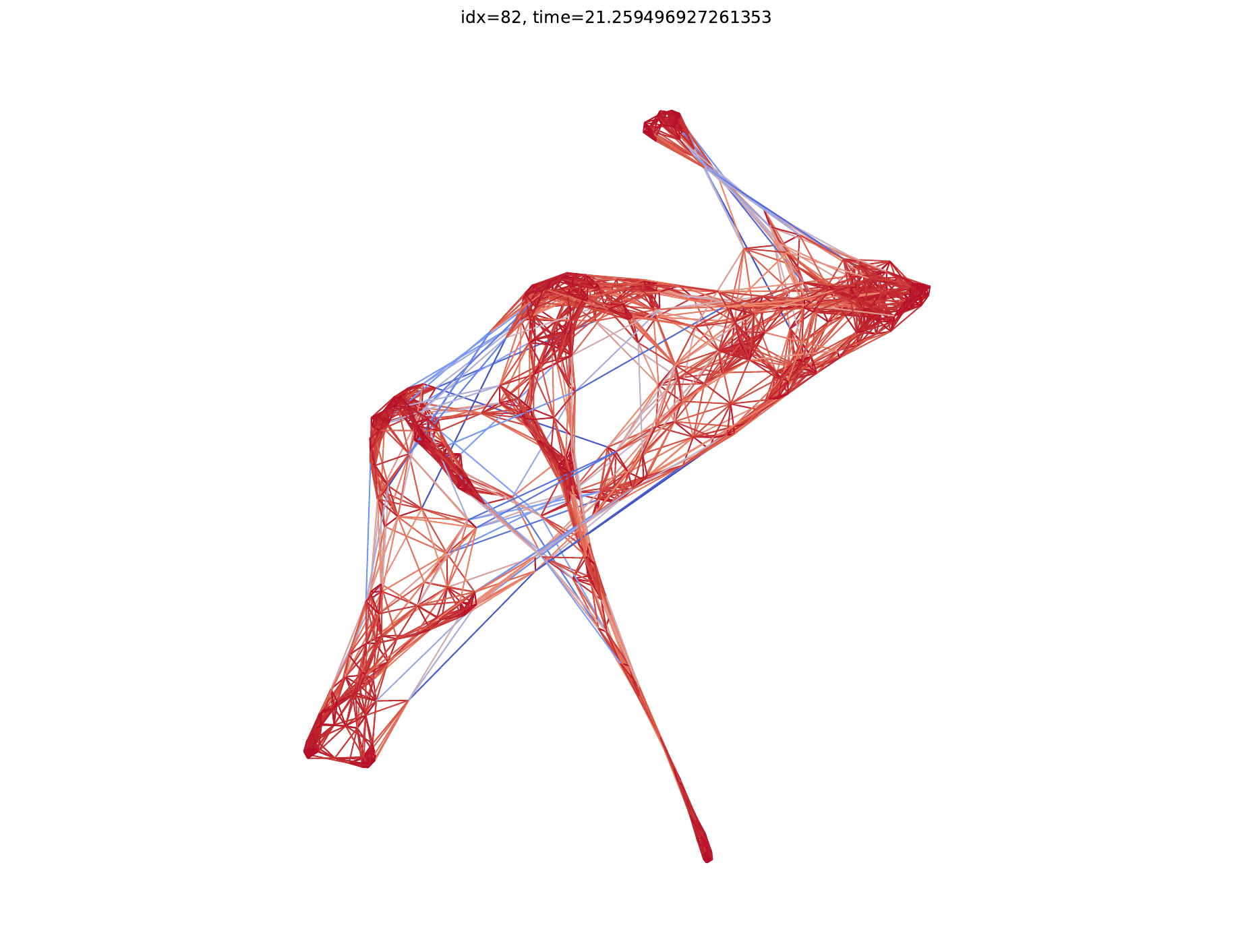} &
\imgcell{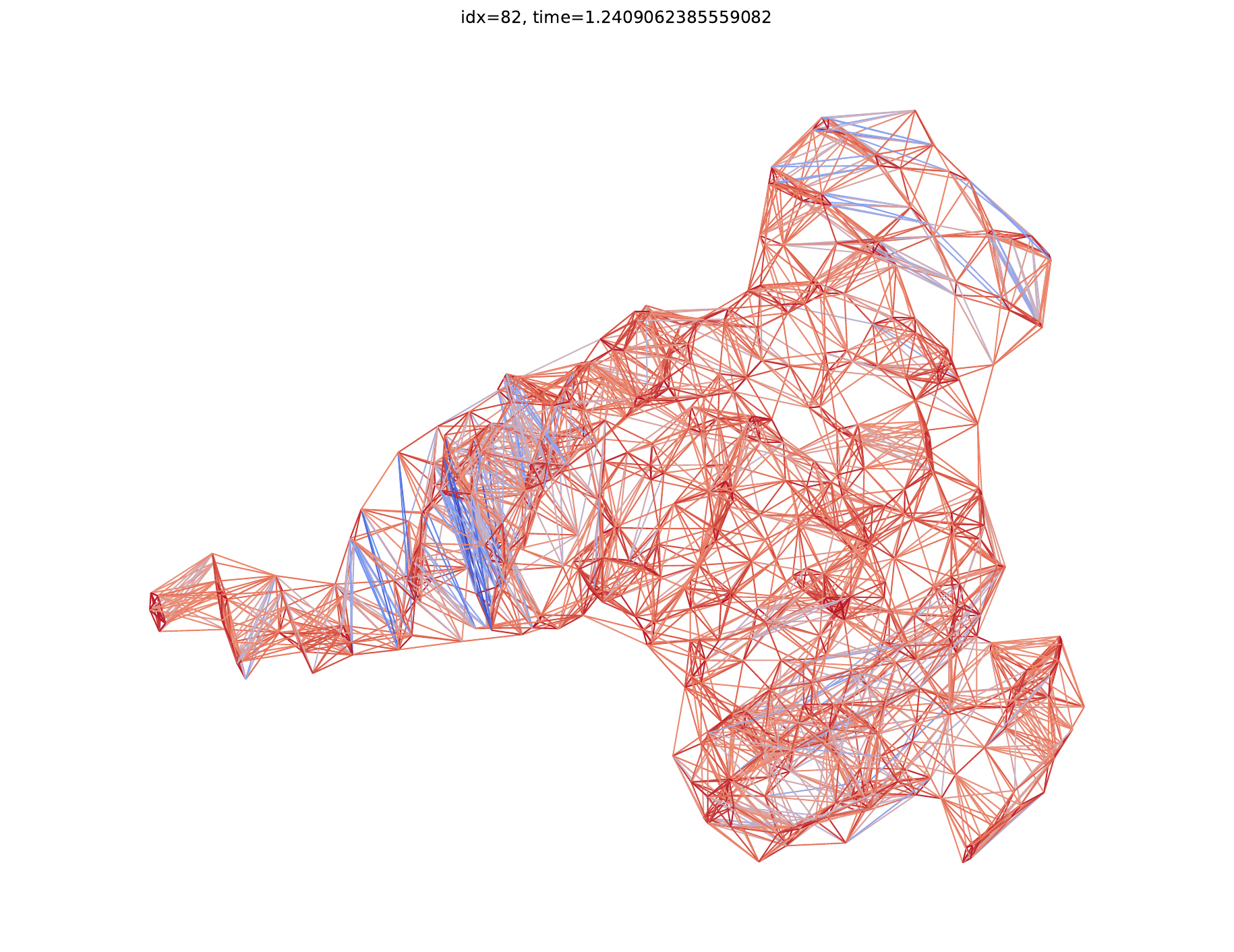} &
\imgcell{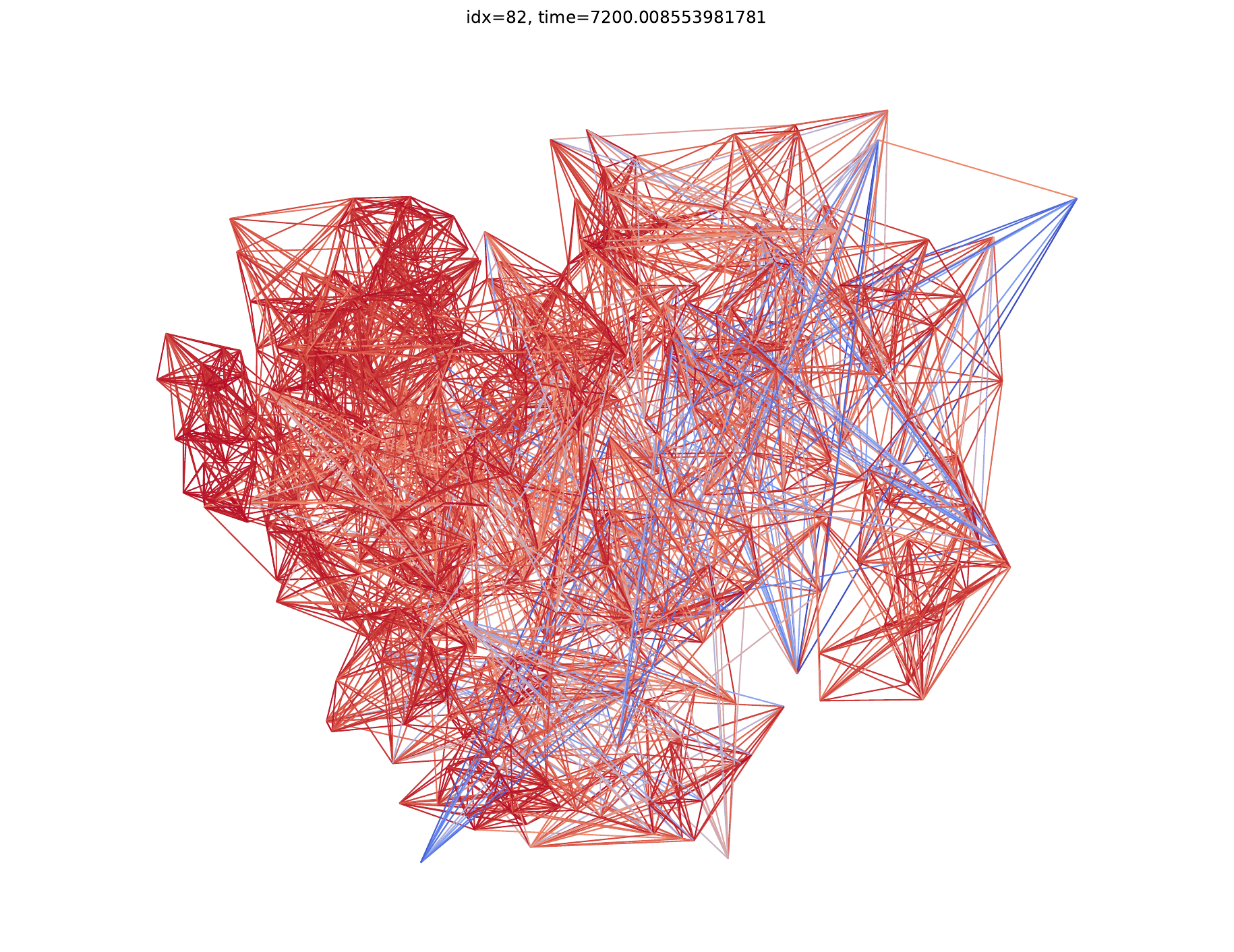} &
\imgcell{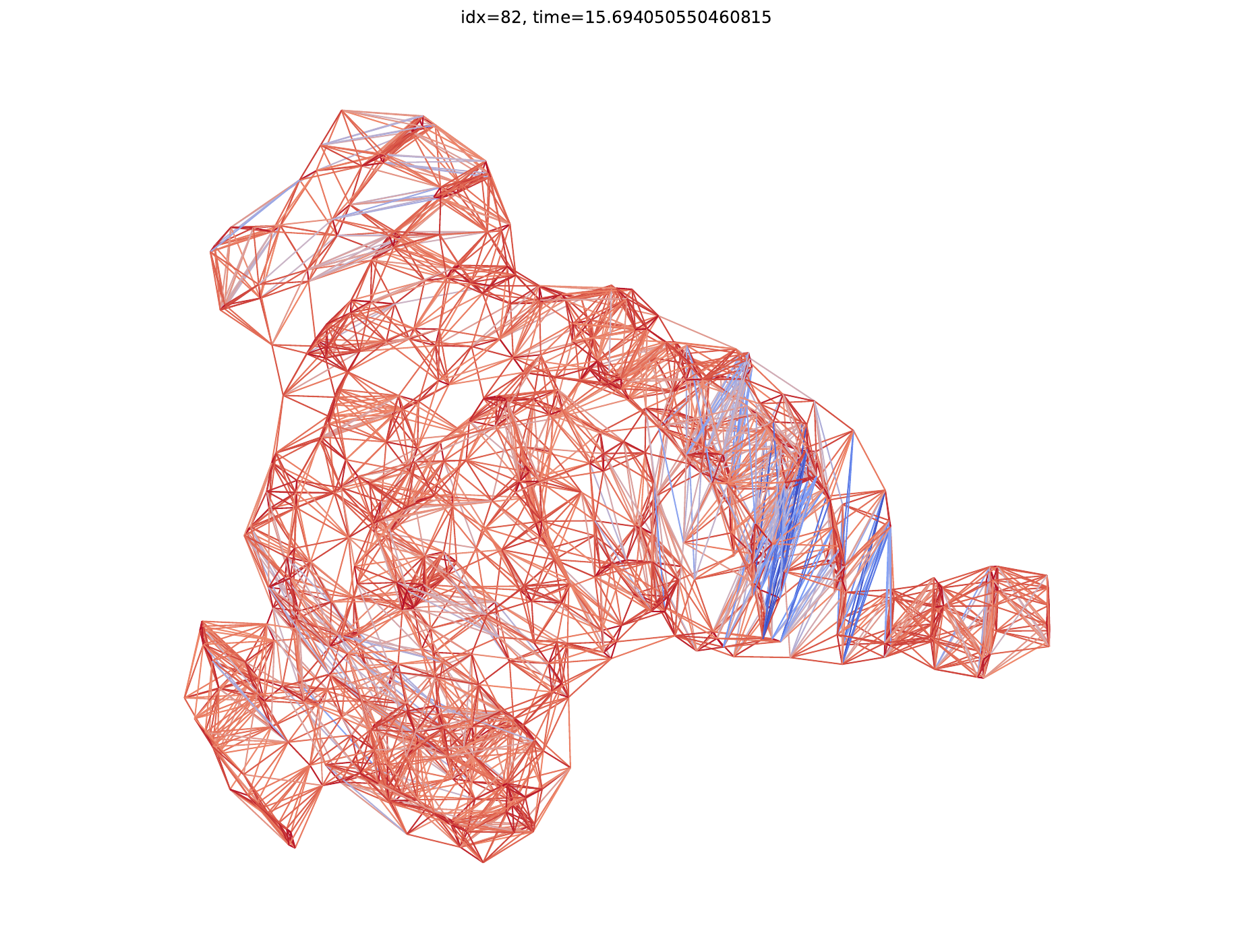} &
\imgcell{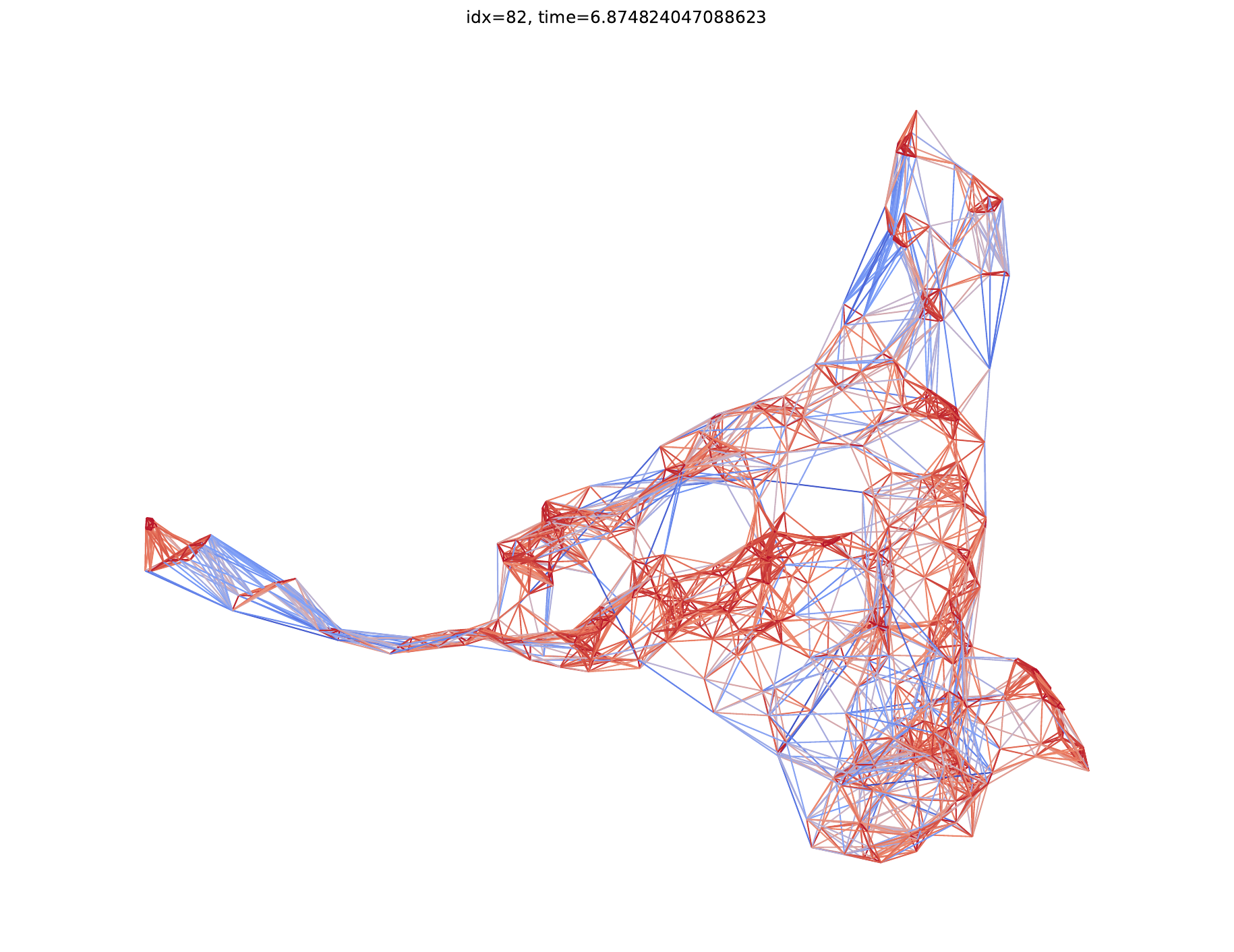} &
\imgcell{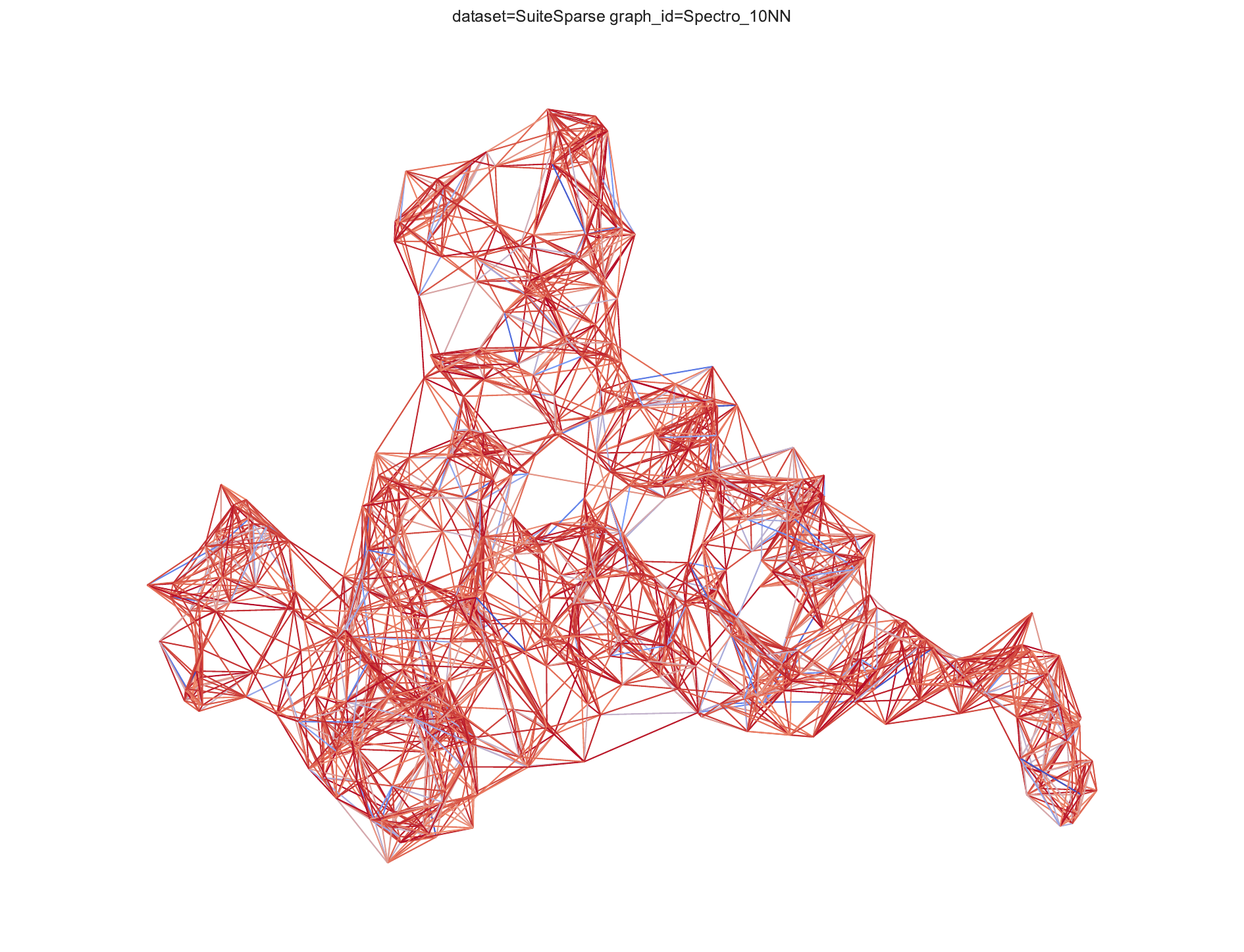} &
\imgcell{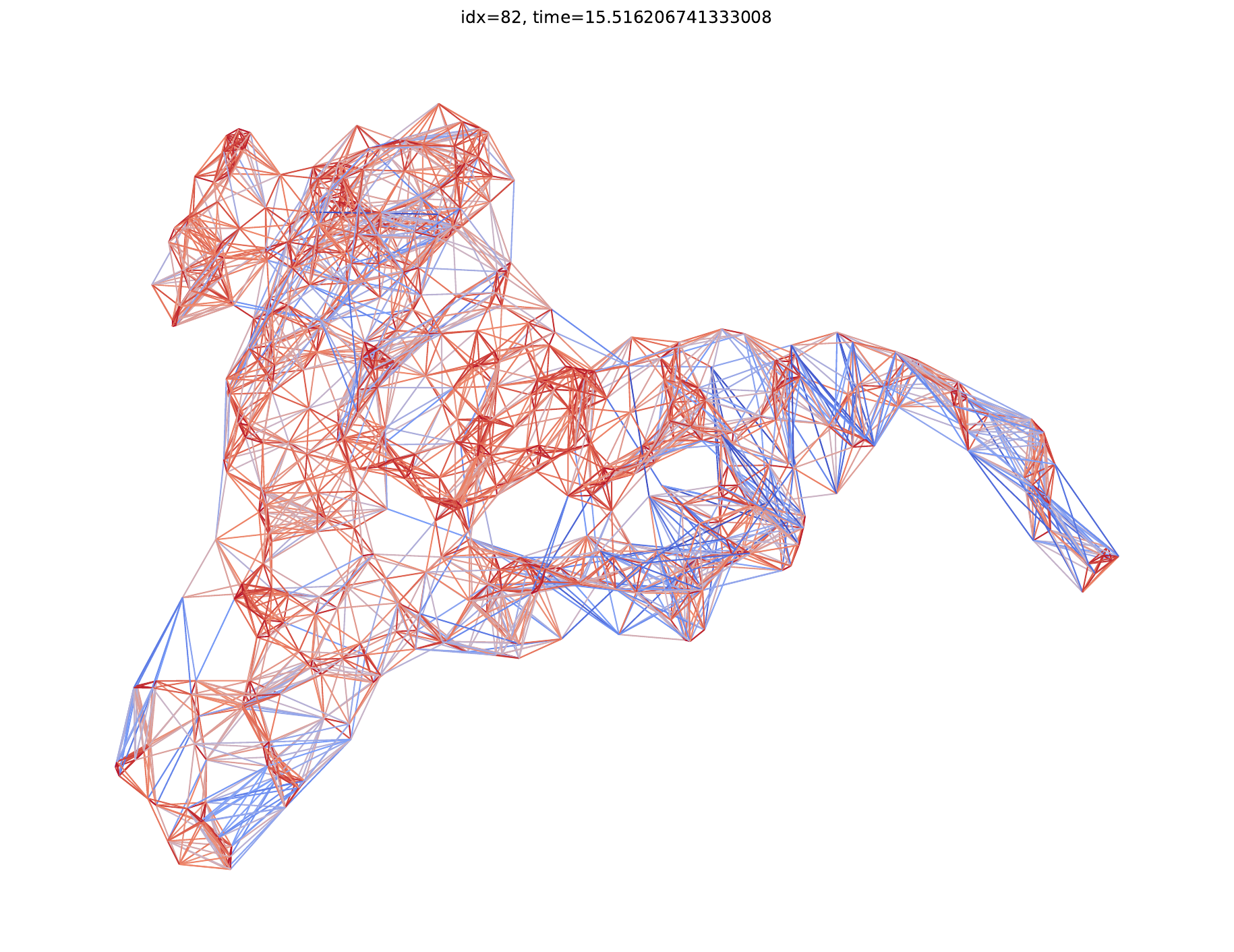} &
\imgcell{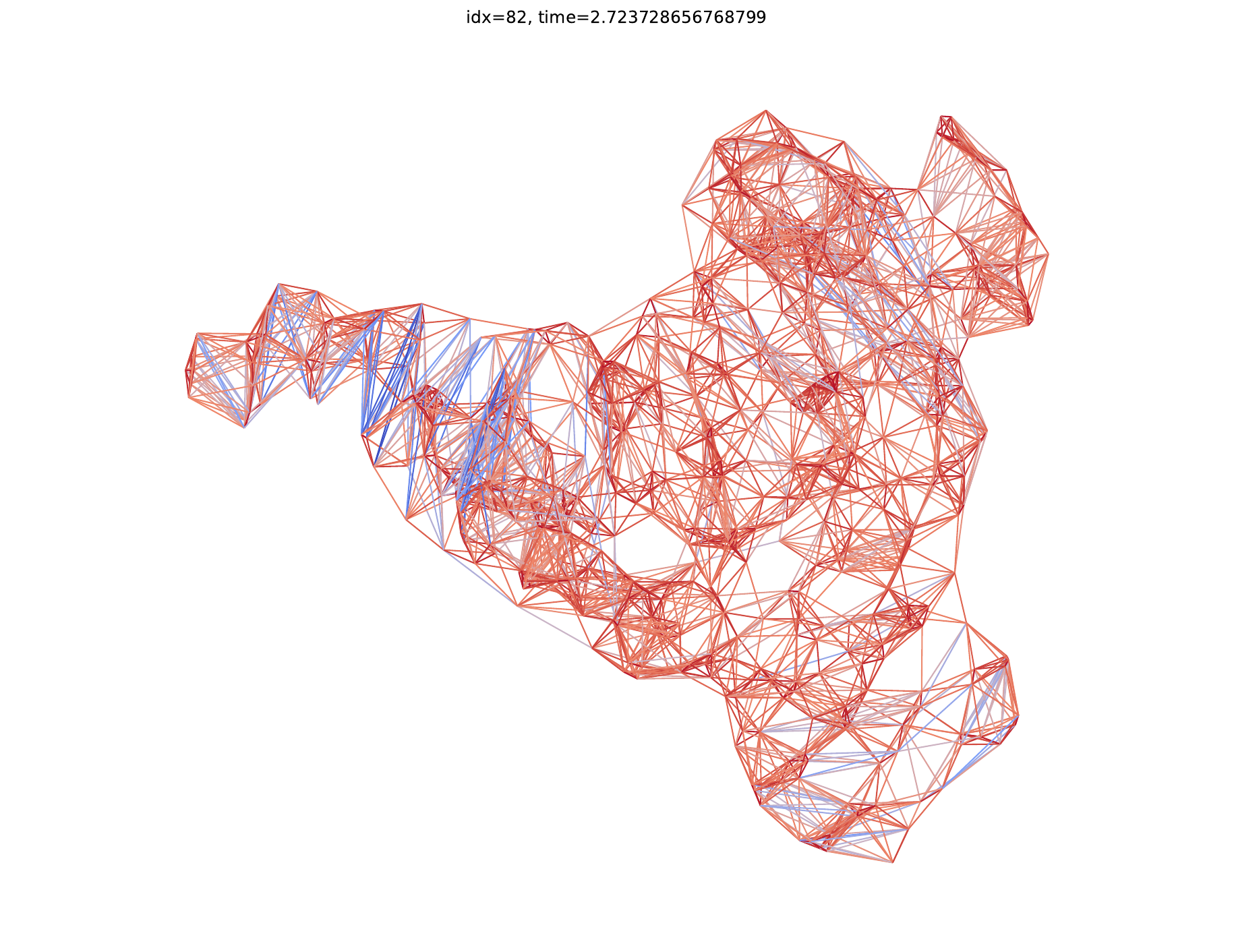} &
\imgcell{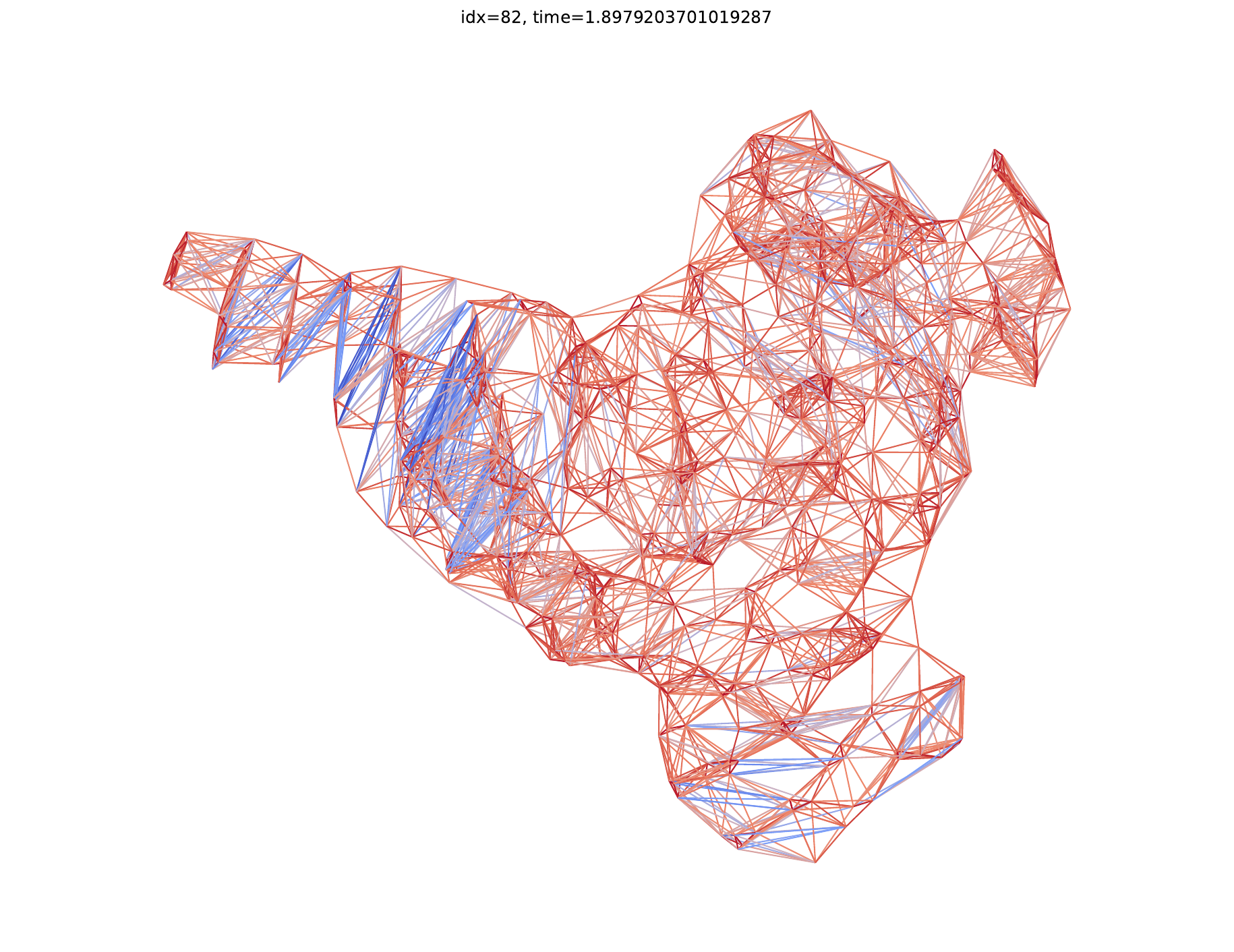} &
\imgcell{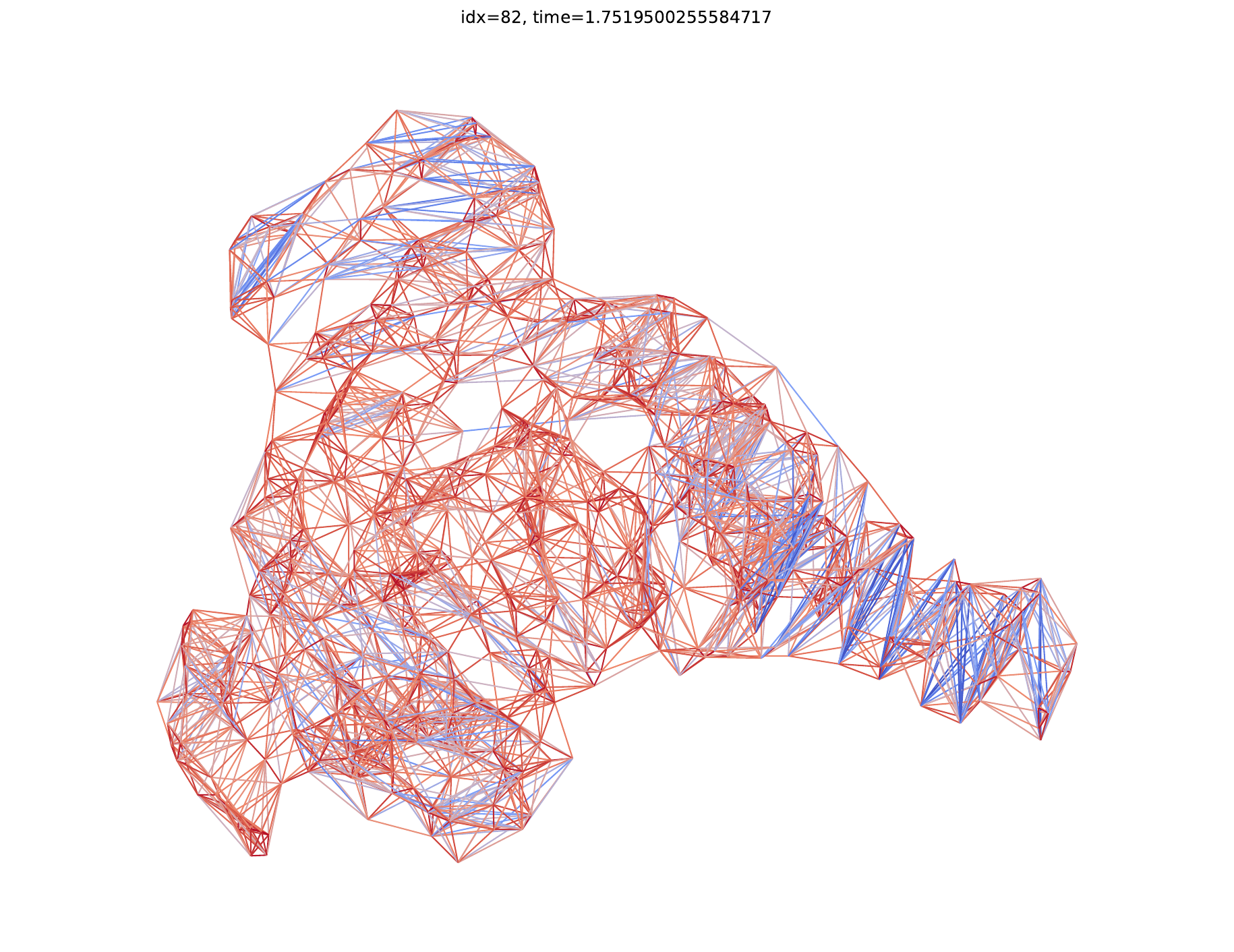} \\

&
t = 0.11s &
t = 14.68s &
t = 21.26s &
t = 1.24s &
t = 7200.00s &
t = 1.01s &
t = 1.14s &
t = 1.20s &
t = 1.12s &
t = 0.84s &
t = 1.22s &
t = 1.18s \\

\makecell{\bfseries micromass\_10NN\\N = 571\\M = 4834} &
\imgcell{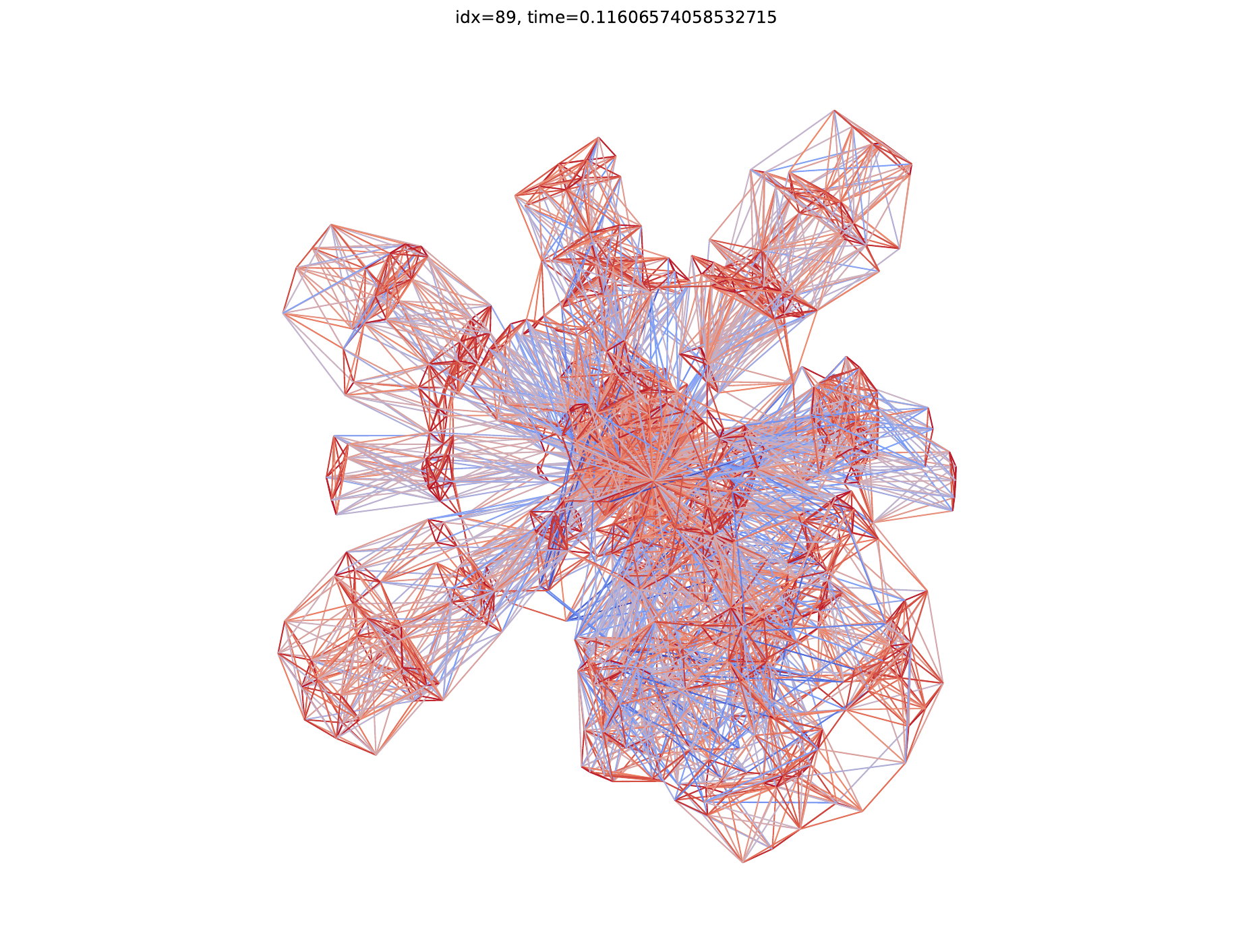} &
\imgcell{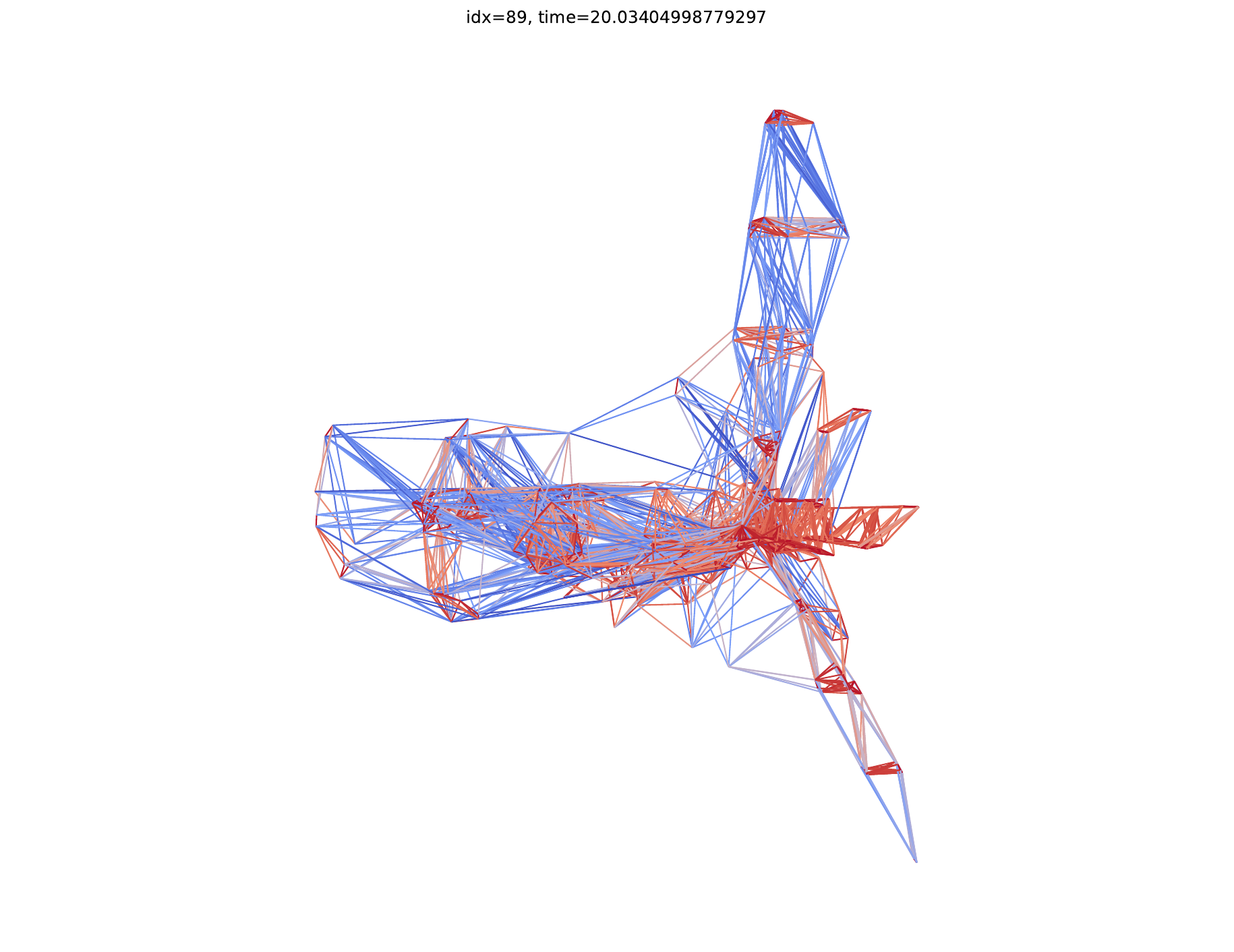} &
\imgcell{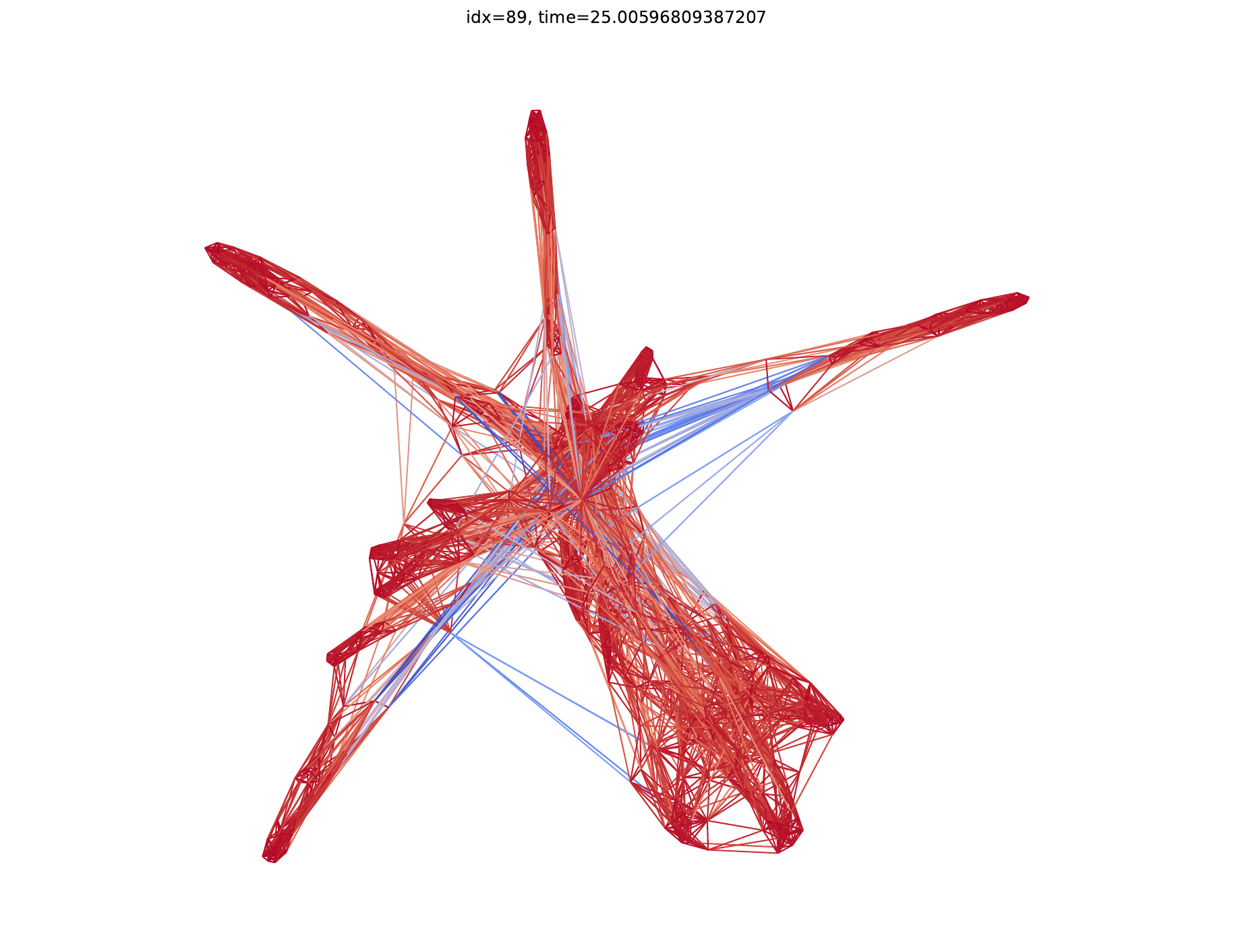} &
\imgcell{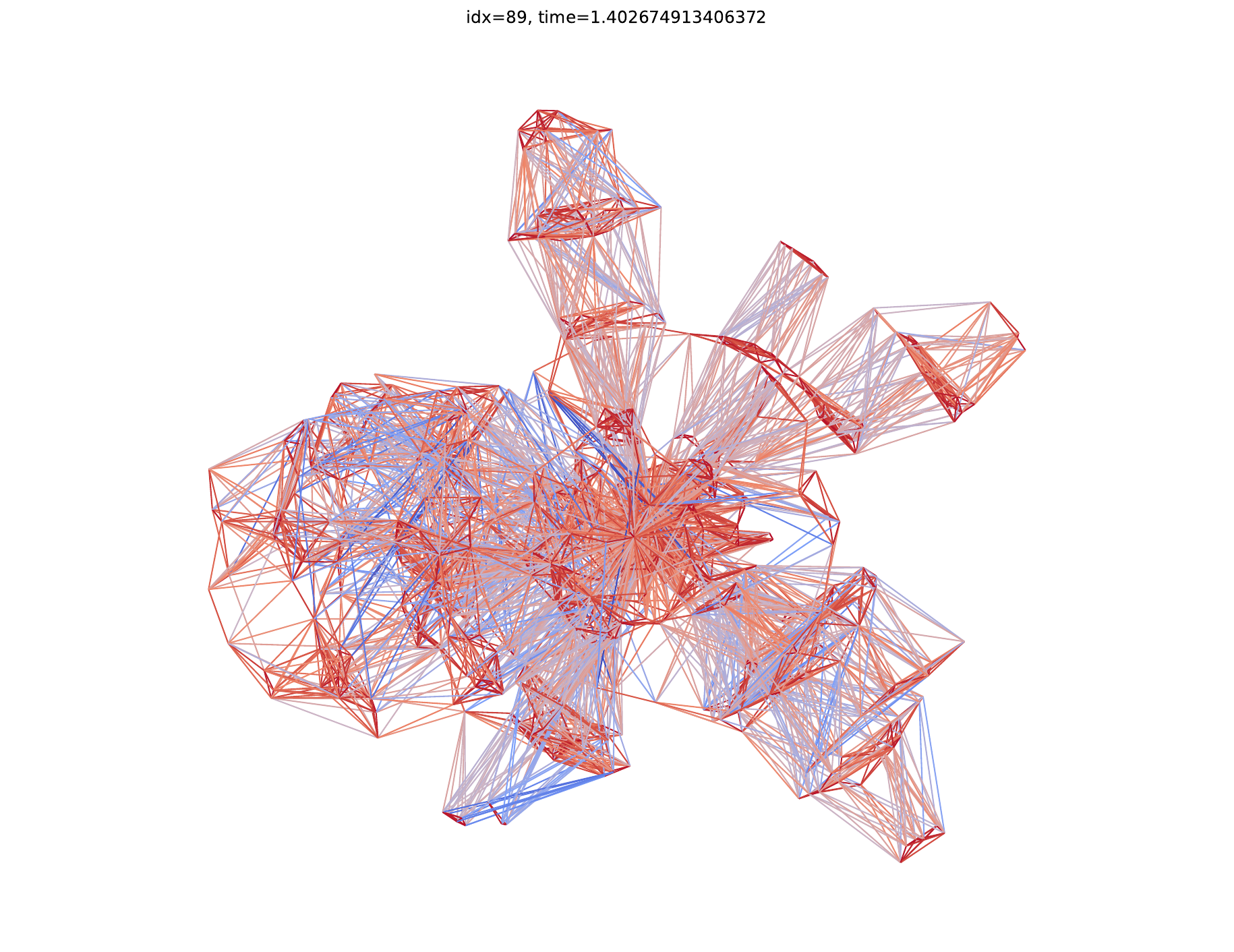} &
\imgcell{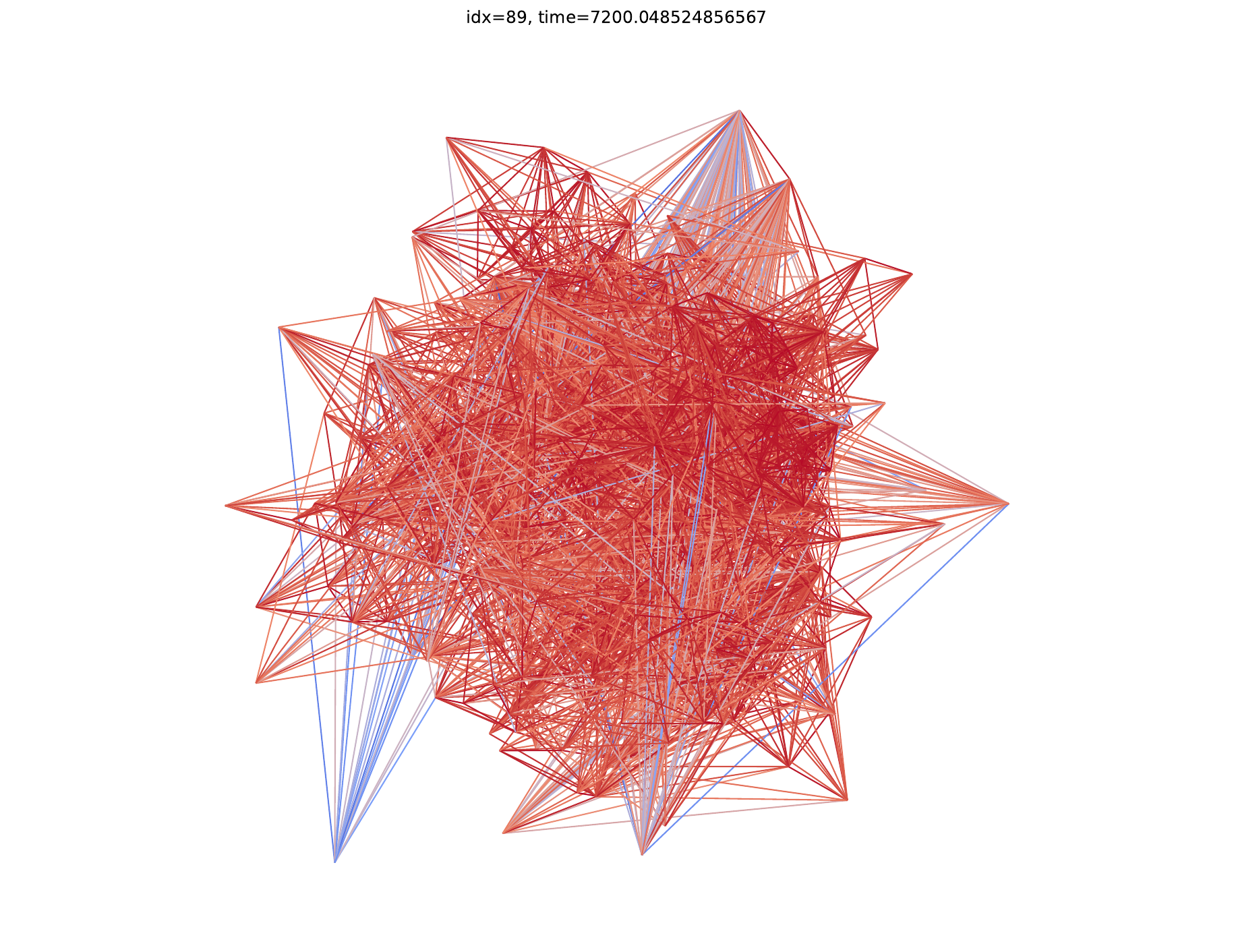} &
\imgcell{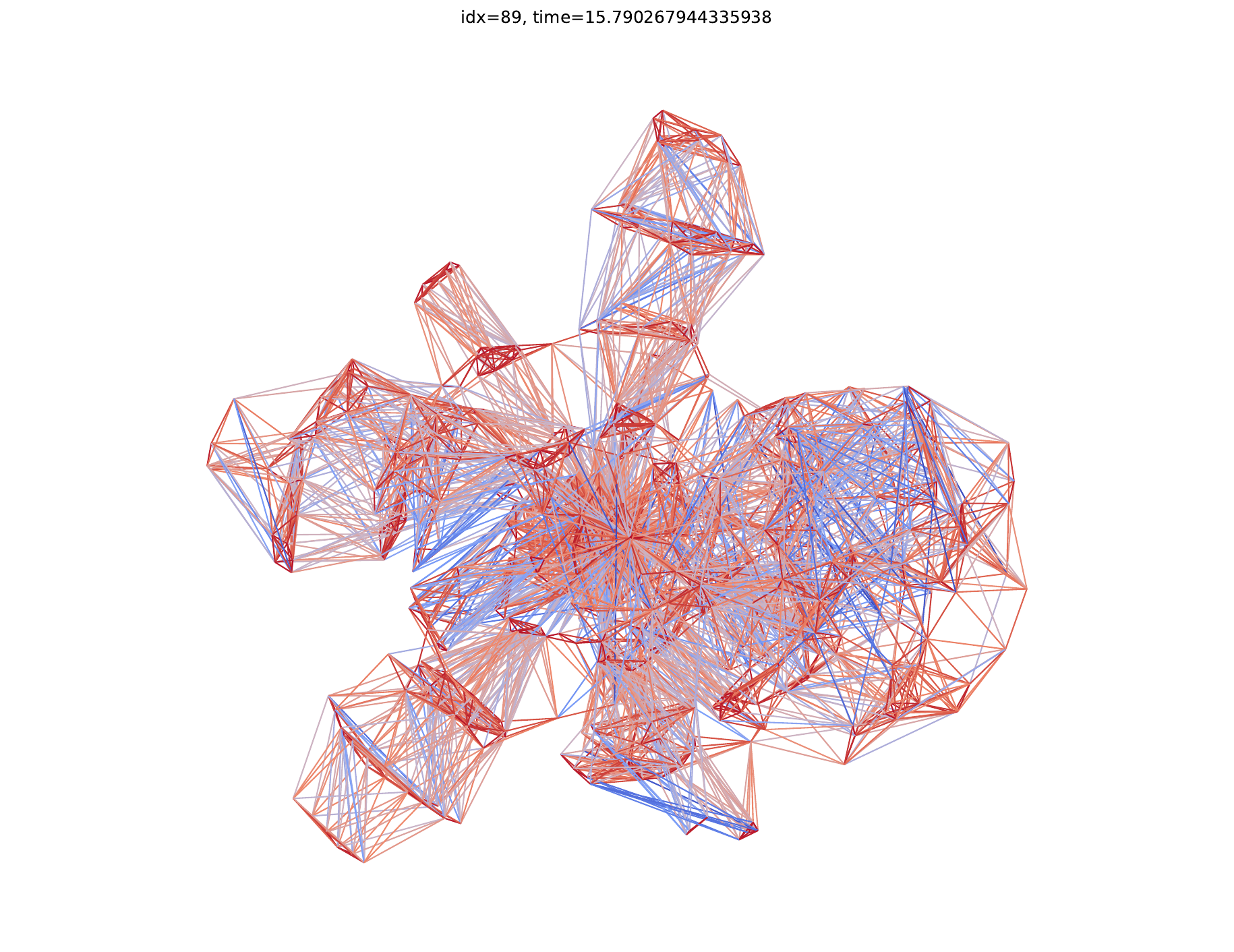} &
\imgcell{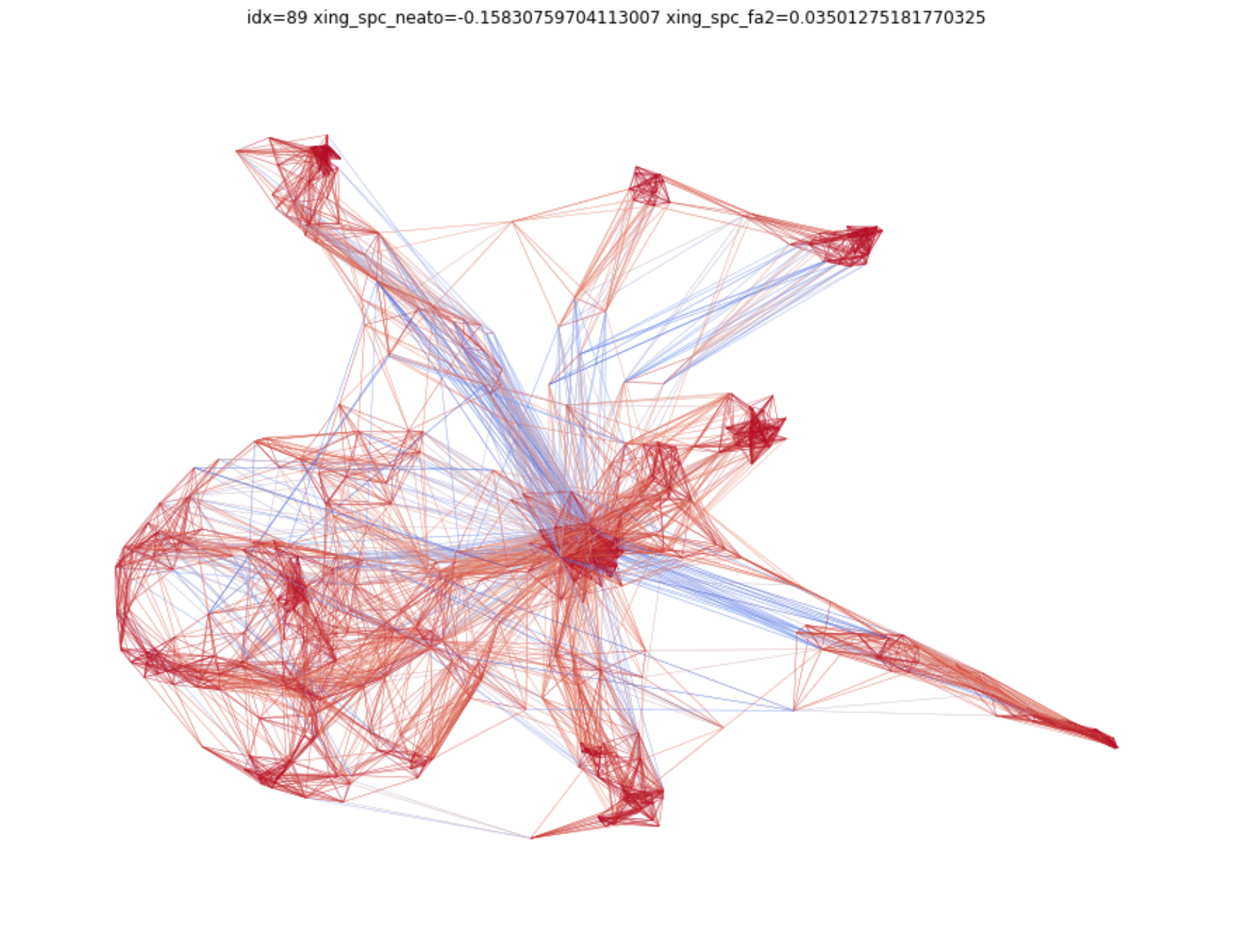} &
\imgcell{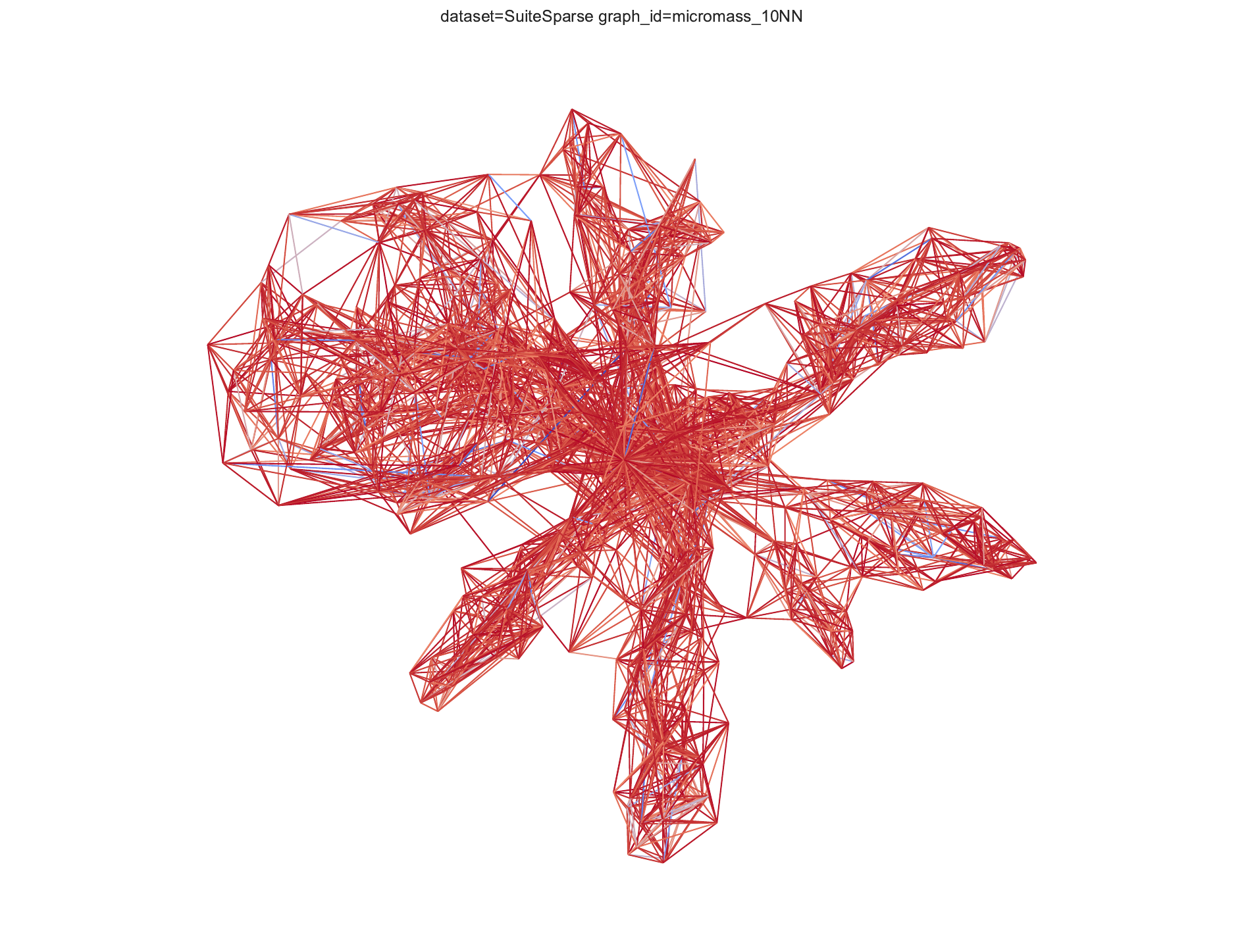} &
\imgcell{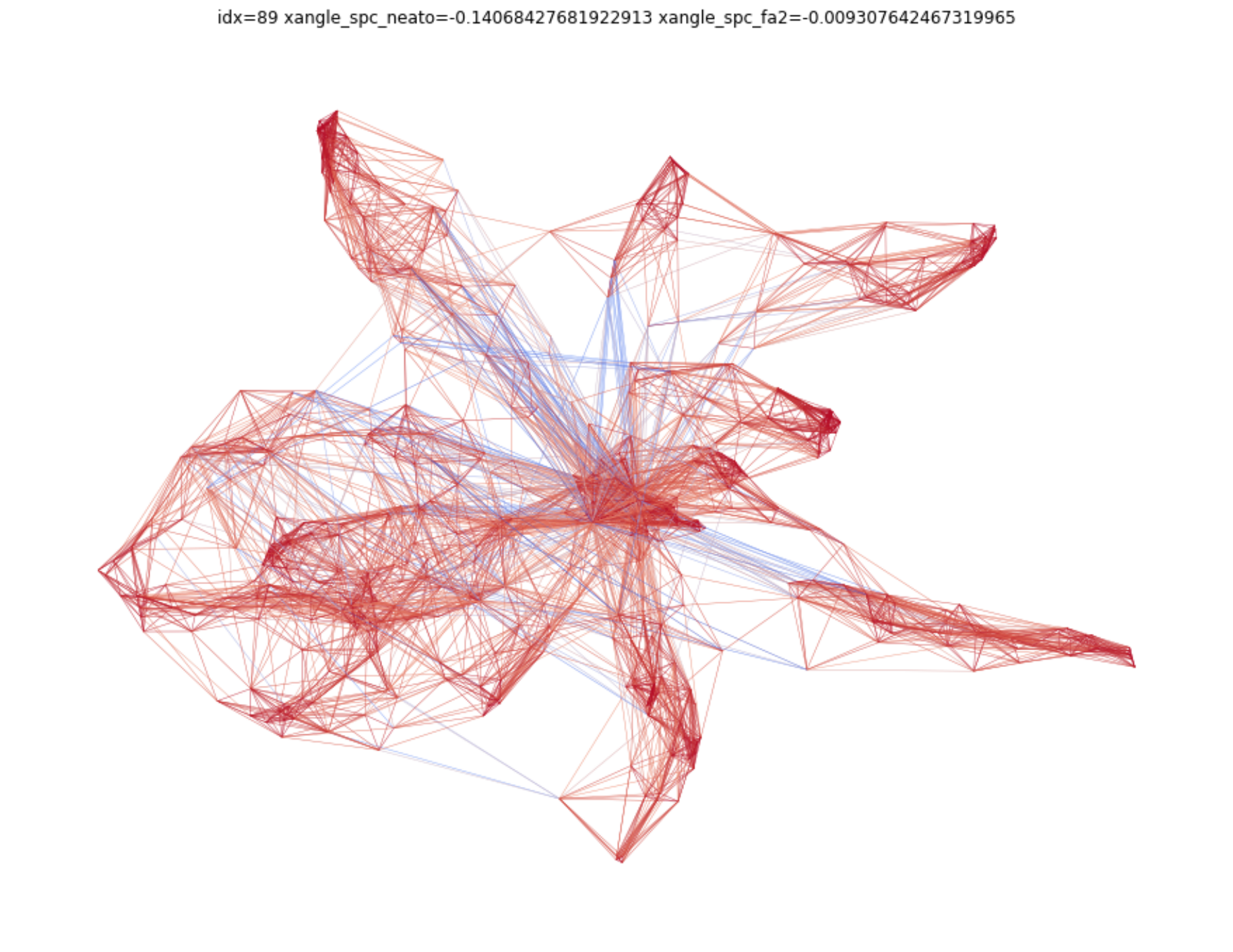} &
\imgcell{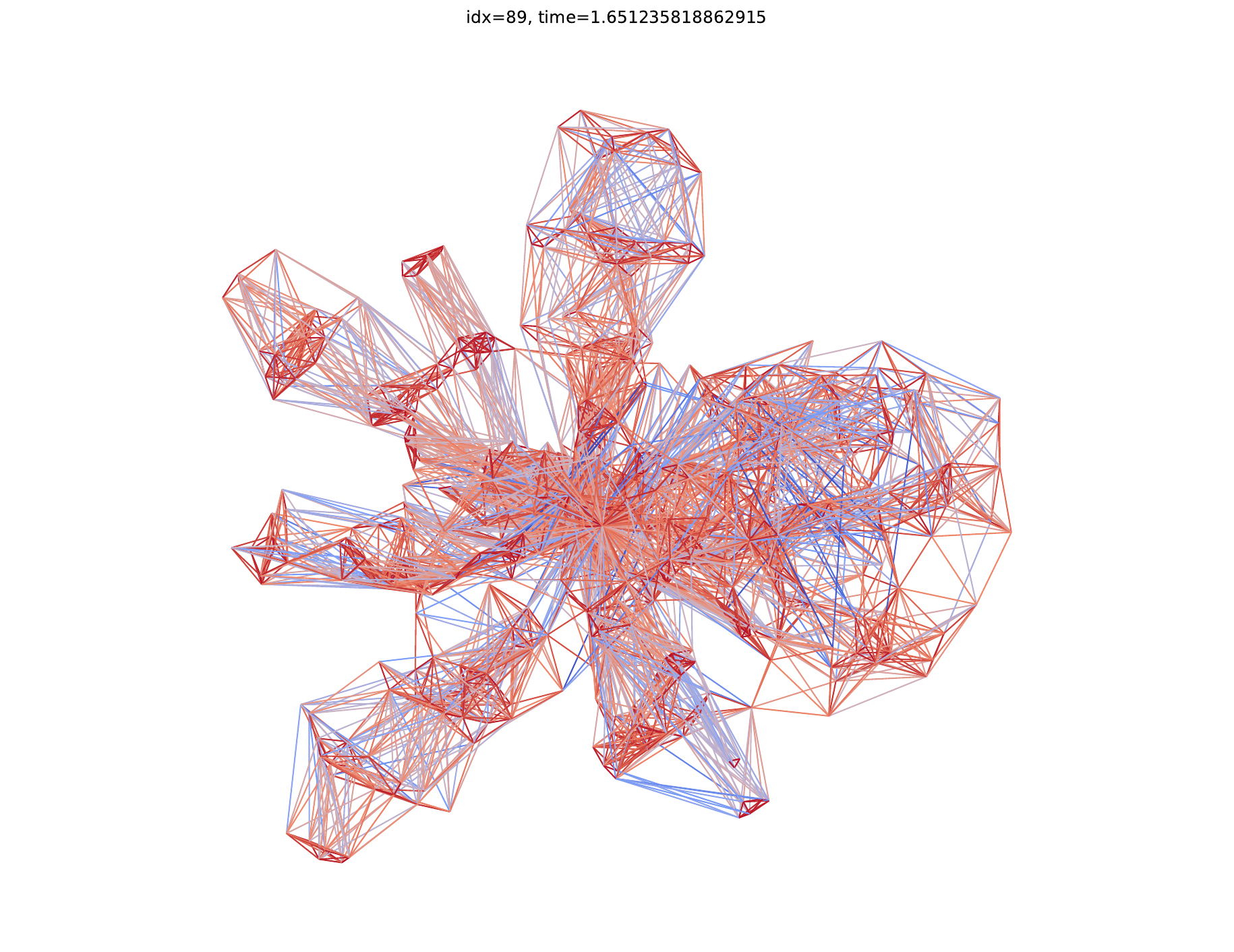} &
\imgcell{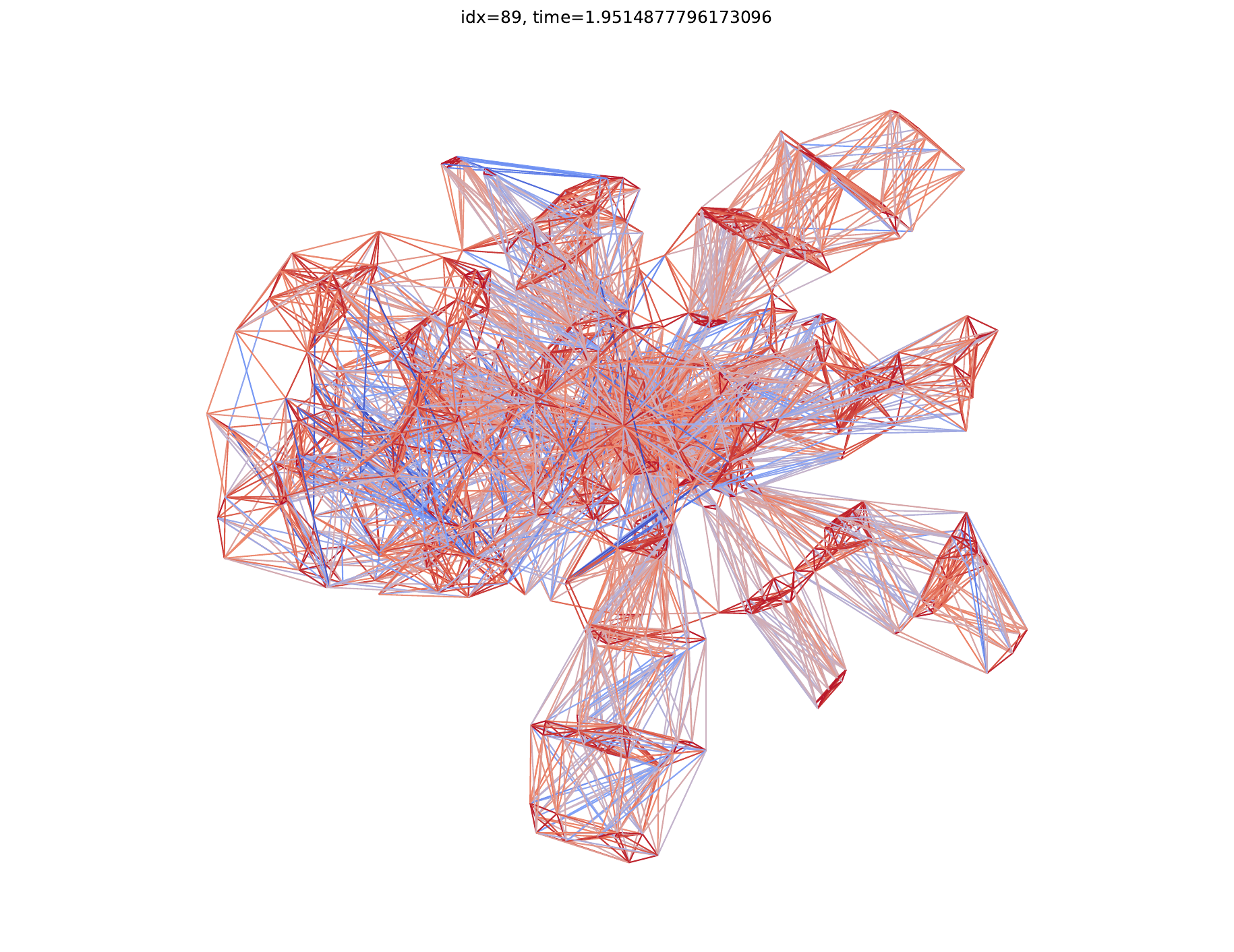} &
\imgcell{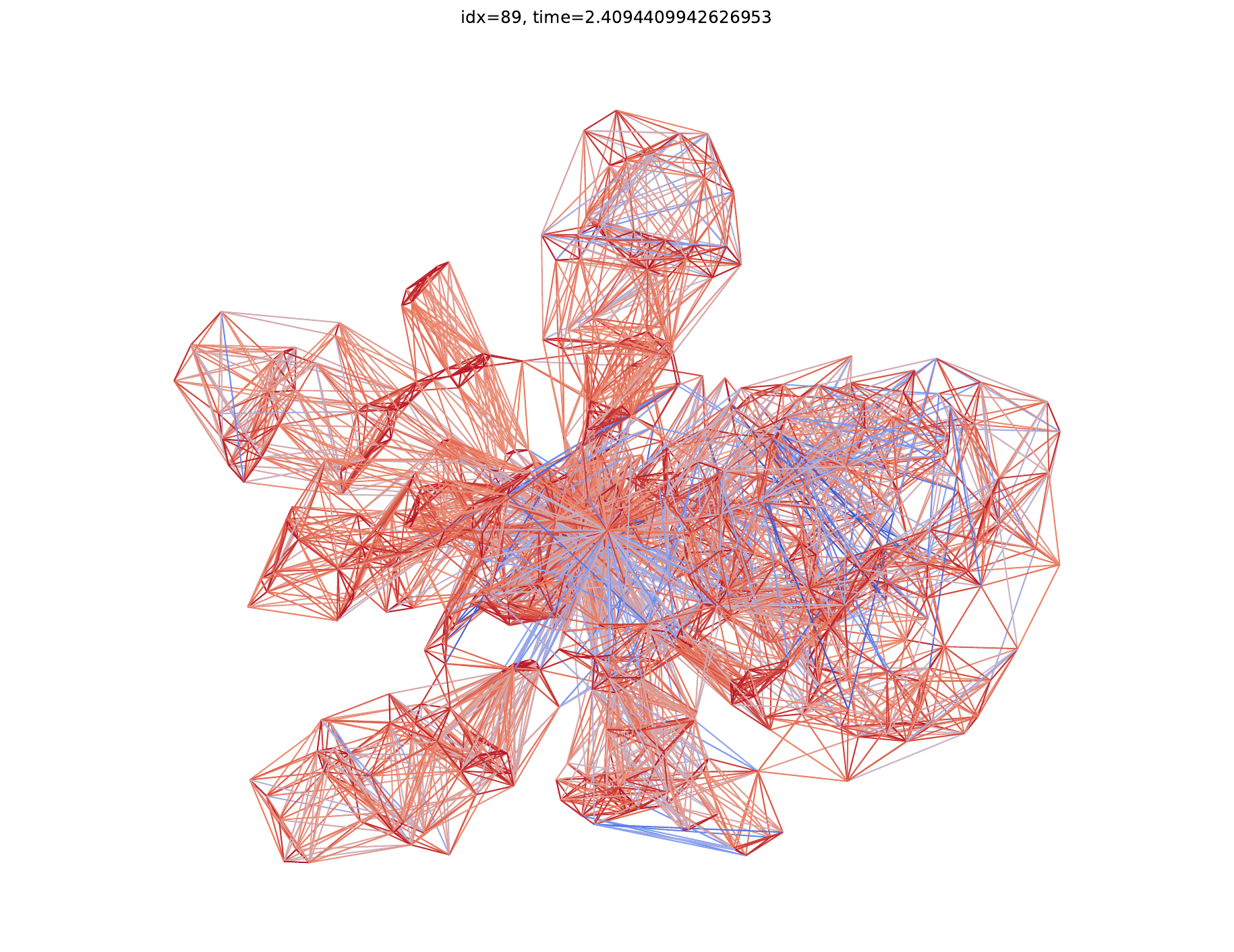} \\

&
t = 0.12s &
t = 20.03s &
t = 25.01s &
t = 1.40s &
t = 7200.00s &
t = 1.16s &
t = 1.26s &
t = 1.47s &
t = 1.60s &
t = 1.34s &
t = 1.51s &
t = 1.32s \\

\end{tabular}
\captionof{figure}[]{The qualitative evaluation of 7 \modelName\ models by comparing with 5 competitive and representative benchmarks. All the graphs presented above are unseen during the training phase of \modelName. The name of the graphs with the number of nodes $N$ and the number of edges $M$ is presented in the row header. For each layout, the computation time $t$ (without including the pre-processing time) on the CPU is computed and reported in seconds. }
\label{fig:more-vis-result1}
\end{table*}

\begin{table*}[ht!]
\setlength{\tabcolsep}{0pt}
\renewcommand{\arraystretch}{0}
\fontsize{6}{6}\selectfont
\centering
\begin{tabular}{ c|ccccc|ccccccc }
    \bfseries{\thead{Graph}} & \multicolumn{5}{c|}{\thead{Benchmark Methods}} & \multicolumn{6}{c}{\thead{SmartGD}}\\
    & \bfseries{SGD2}
    & \bfseries{PMDS}
    & \bfseries{FA2}
    & \bfseries{DeepGD}
    & \makecell{\bfseries GD2\\\relax[Stress+Xing]}
    & \makecell{\bfseries SmartGD\\\relax[Stress]}
    & \makecell{\bfseries SmartGD\\\relax[Xing]}
    & \makecell{\bfseries SmartGD\\\relax[Shape]}
    & \makecell{\bfseries SmartGD\\\relax[XAngle]}
    & \makecell{\bfseries SmartGD\\\relax[Stress+Xing]}
    & \makecell{\bfseries SmartGD\\\relax[Stress+XAngle]}
    & \makecell{\bfseries SmartGD\\\relax[7-Aesthetics]}
    \rule[-1ex]{0pt}{0ex} \\ \hline

\makecell{\bfseries bcsstm34\\N = 588\\M = 11841} &
\imgcell{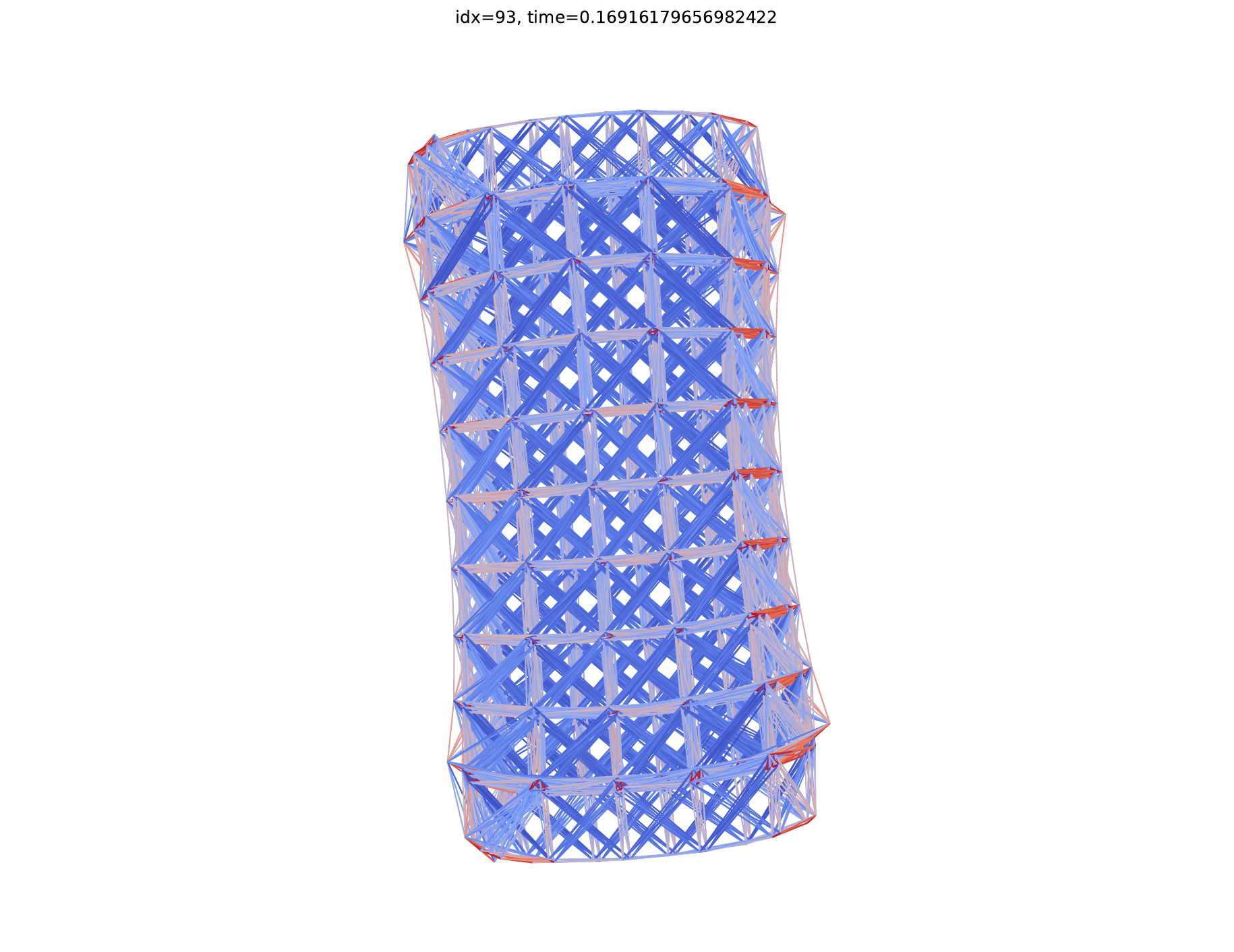} &
\imgcell{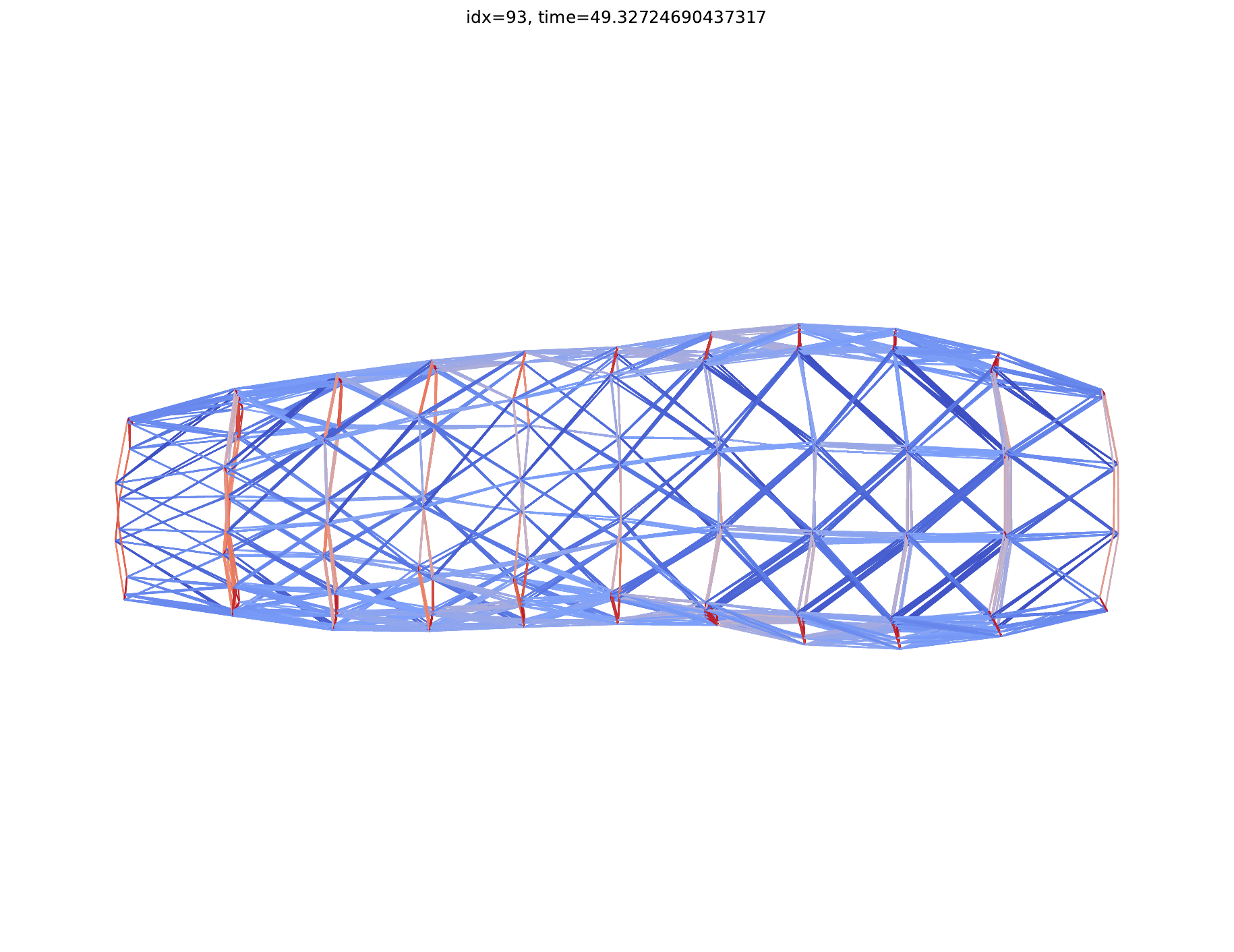} &
\imgcell{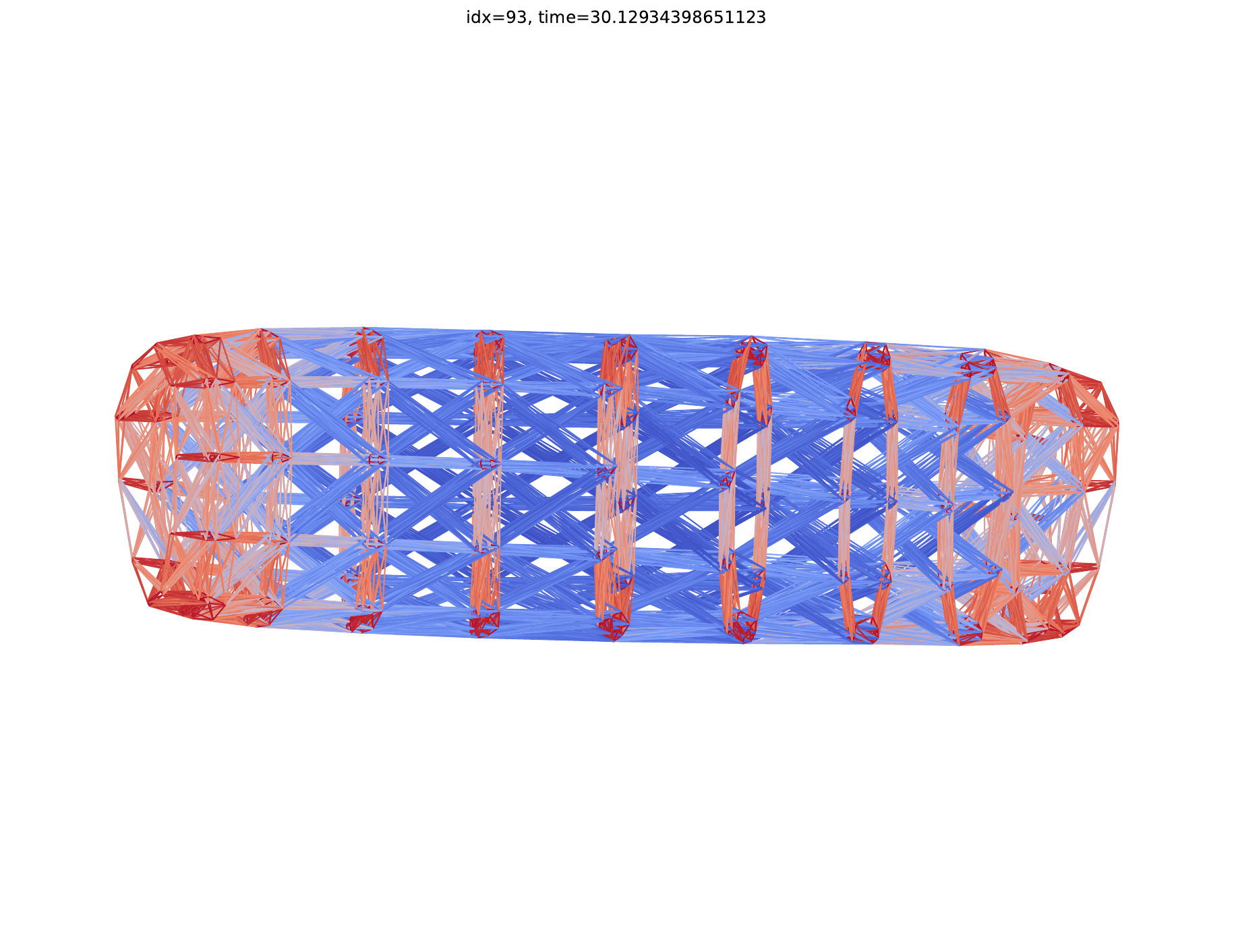} &
\imgcell{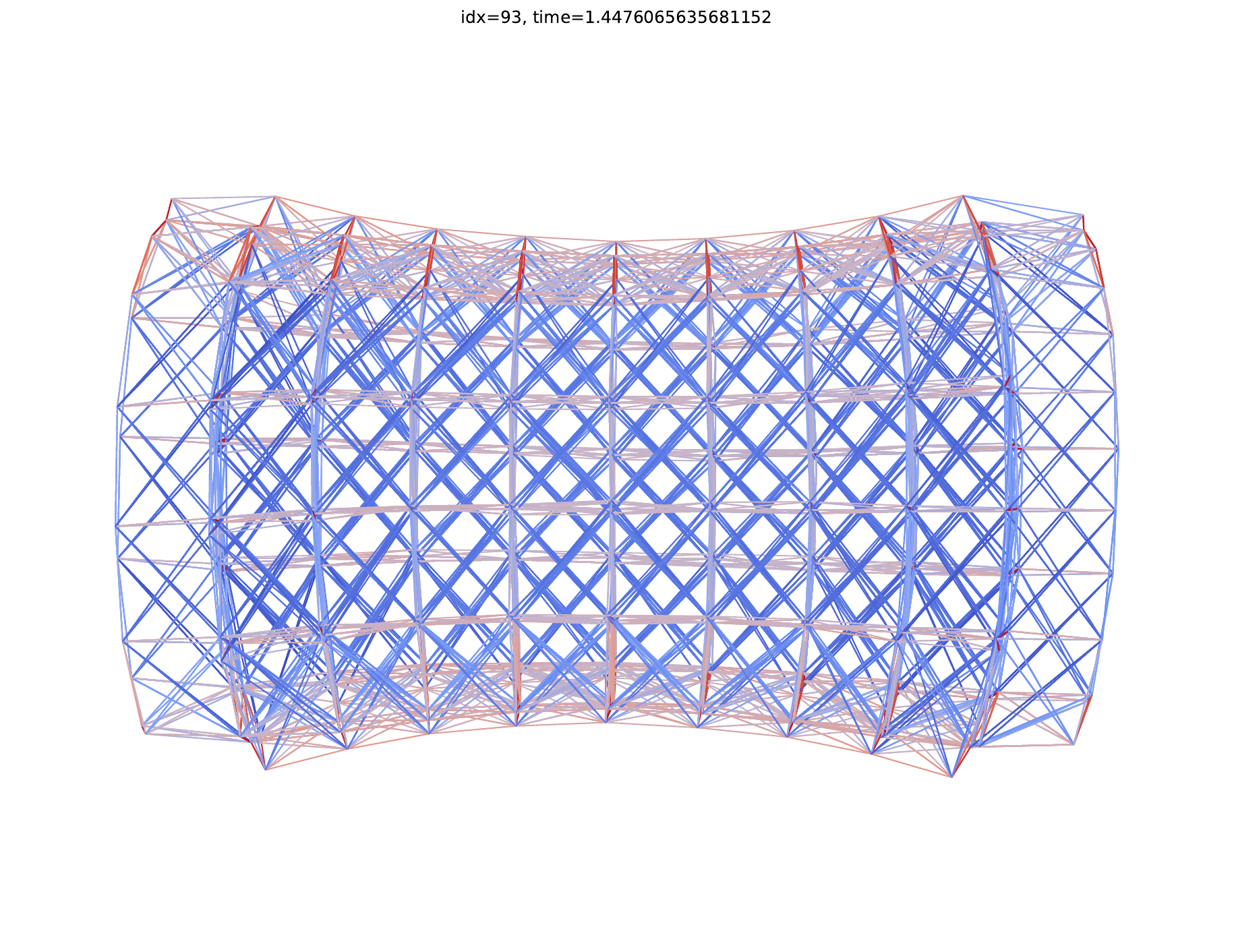} &
\imgcell{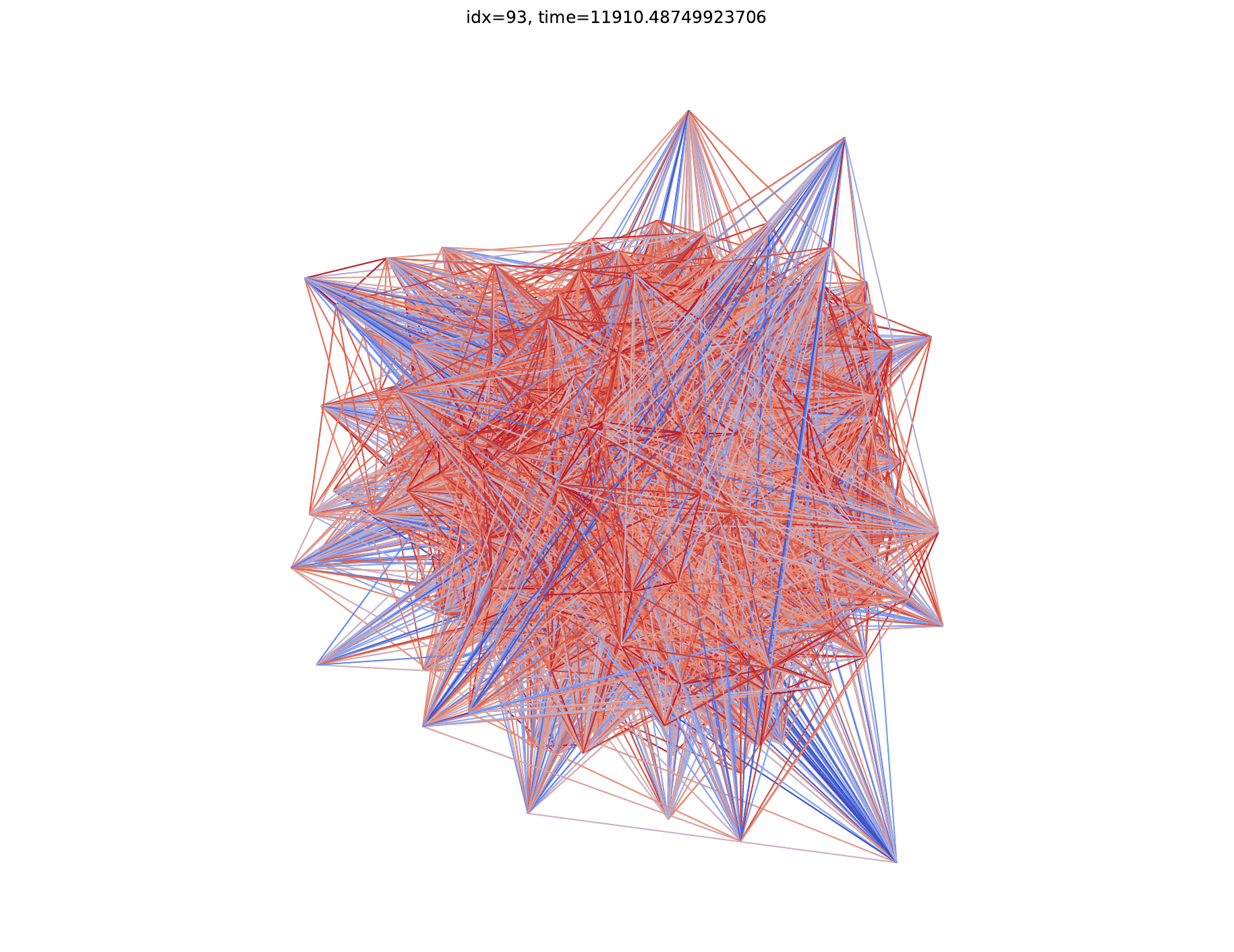} &
\imgcell{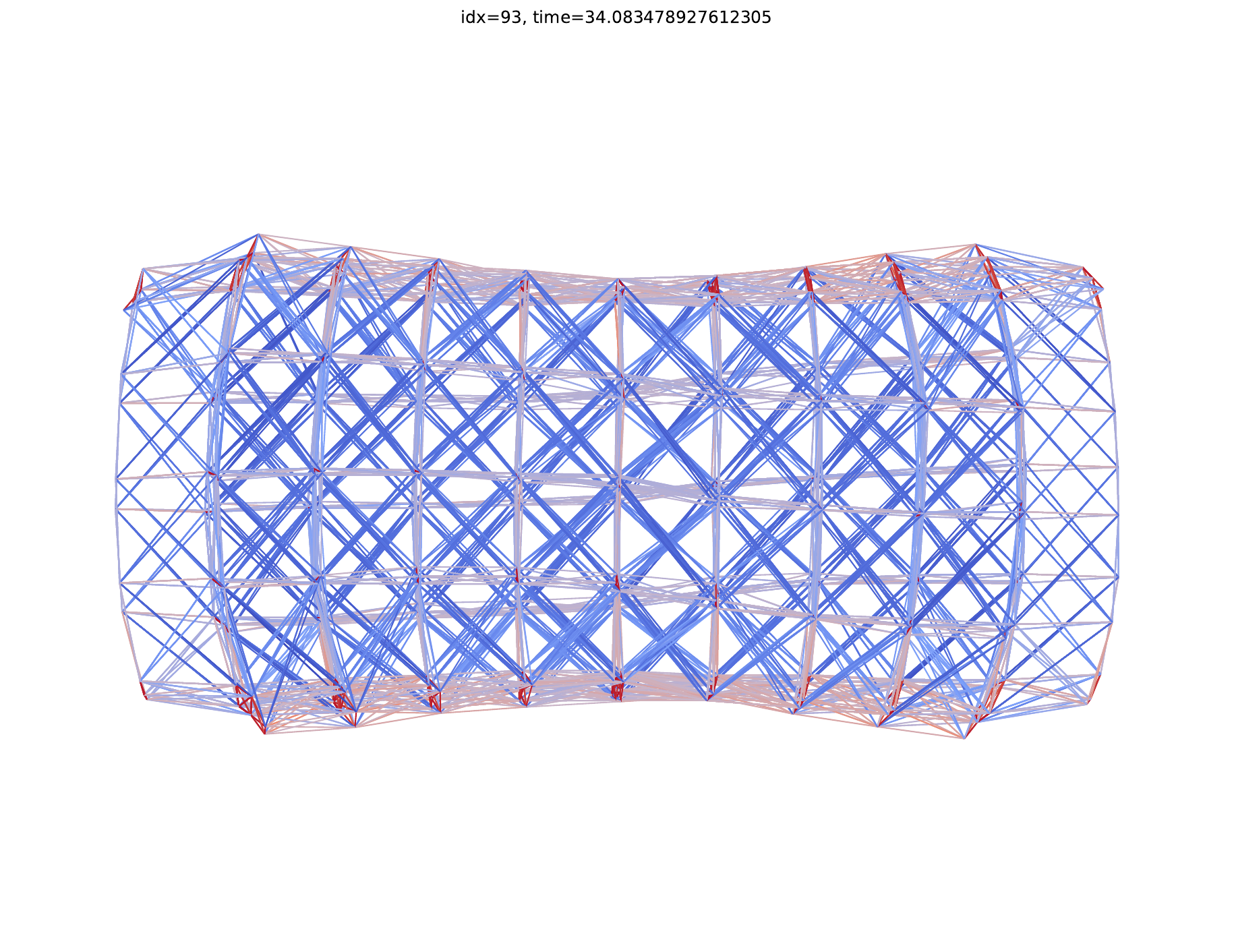} &
\imgcell{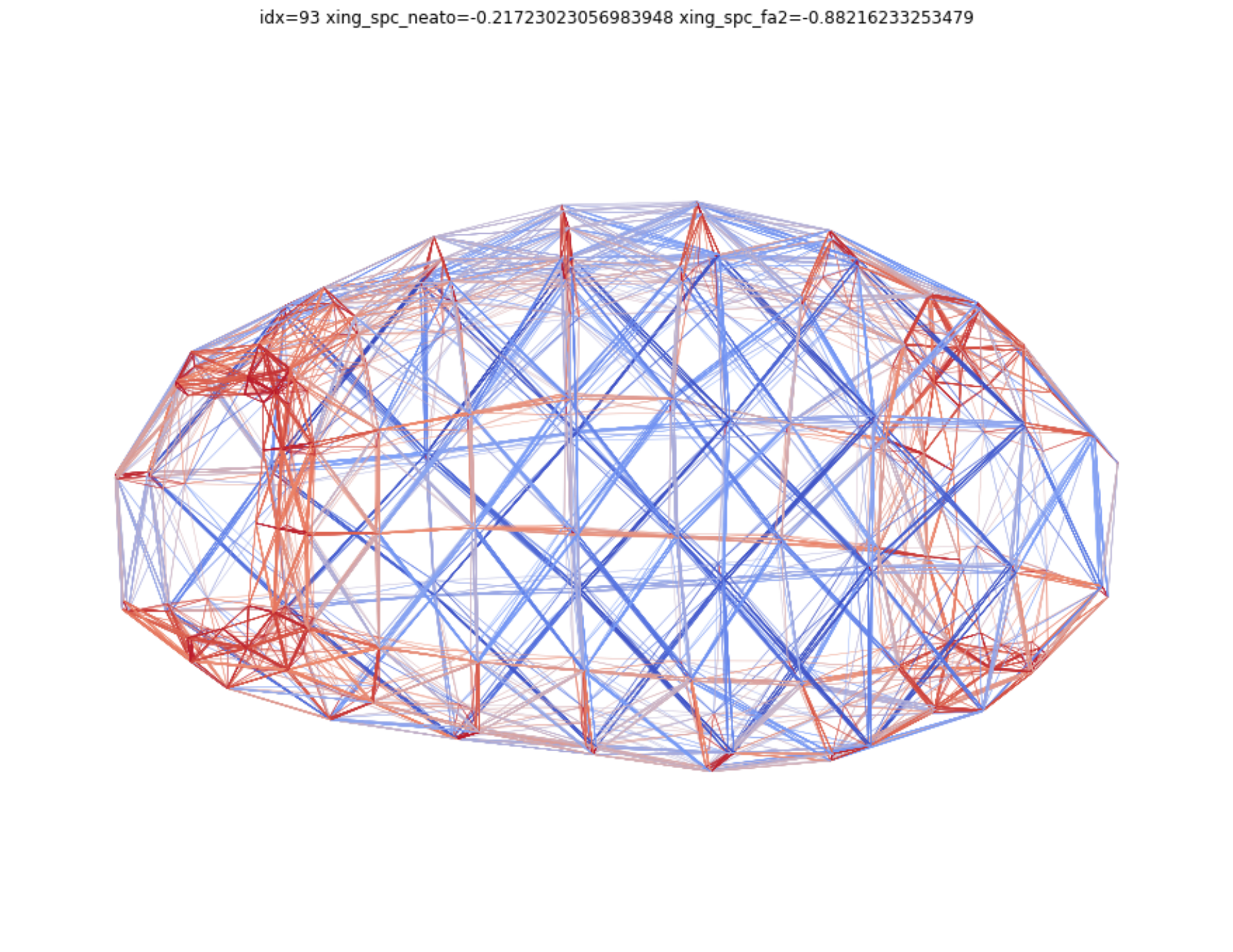} &
\imgcell{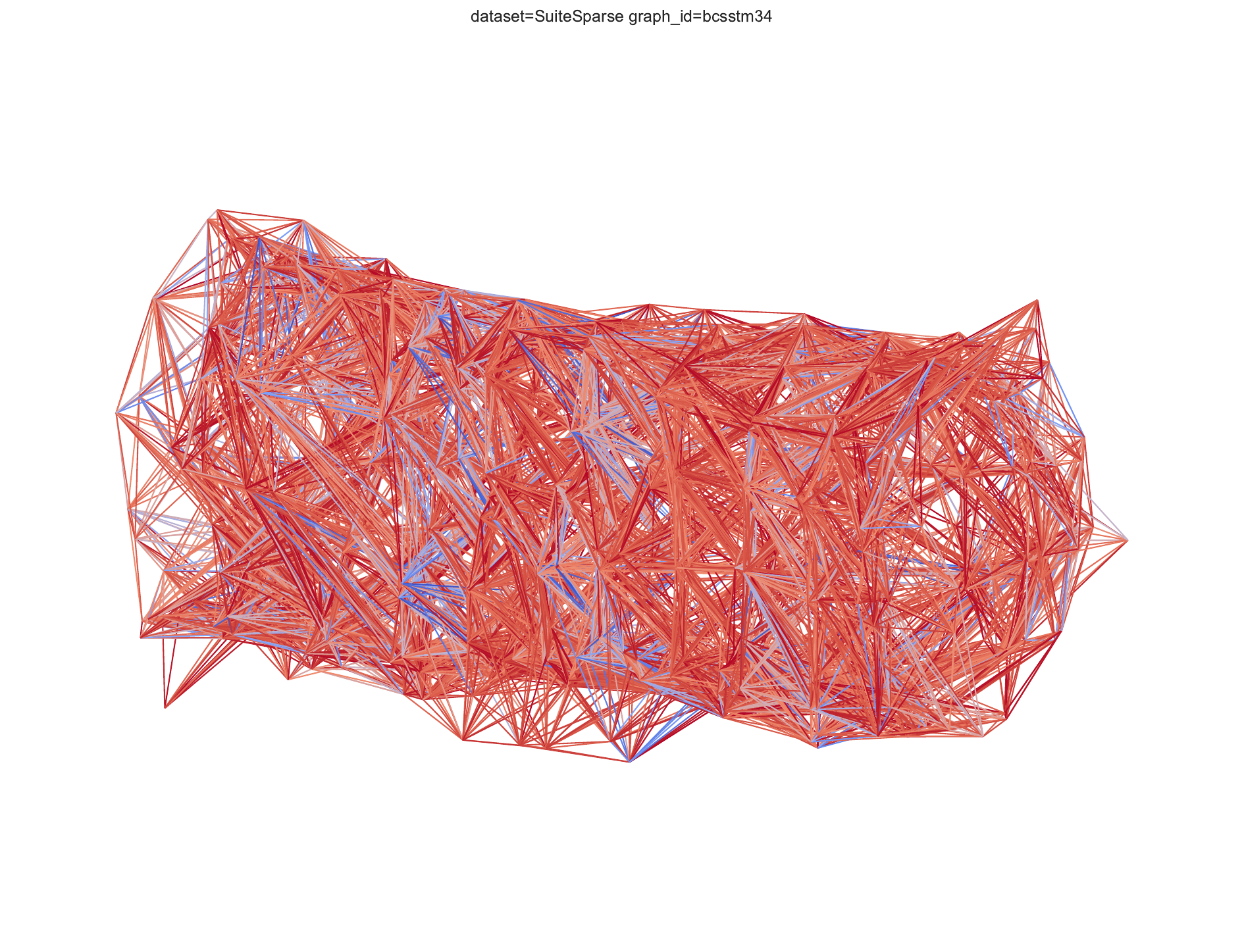} &
\imgcell{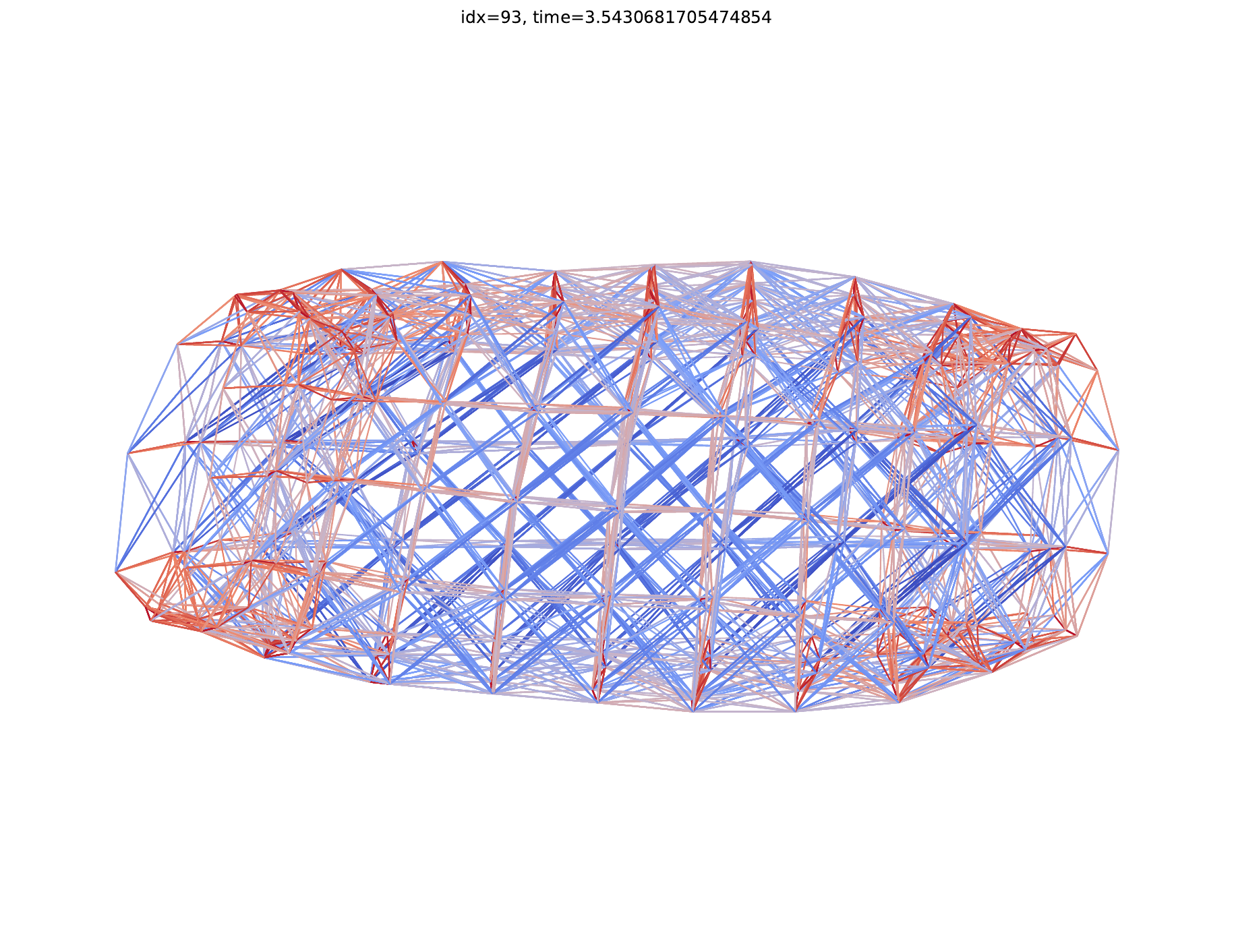} &
\imgcell{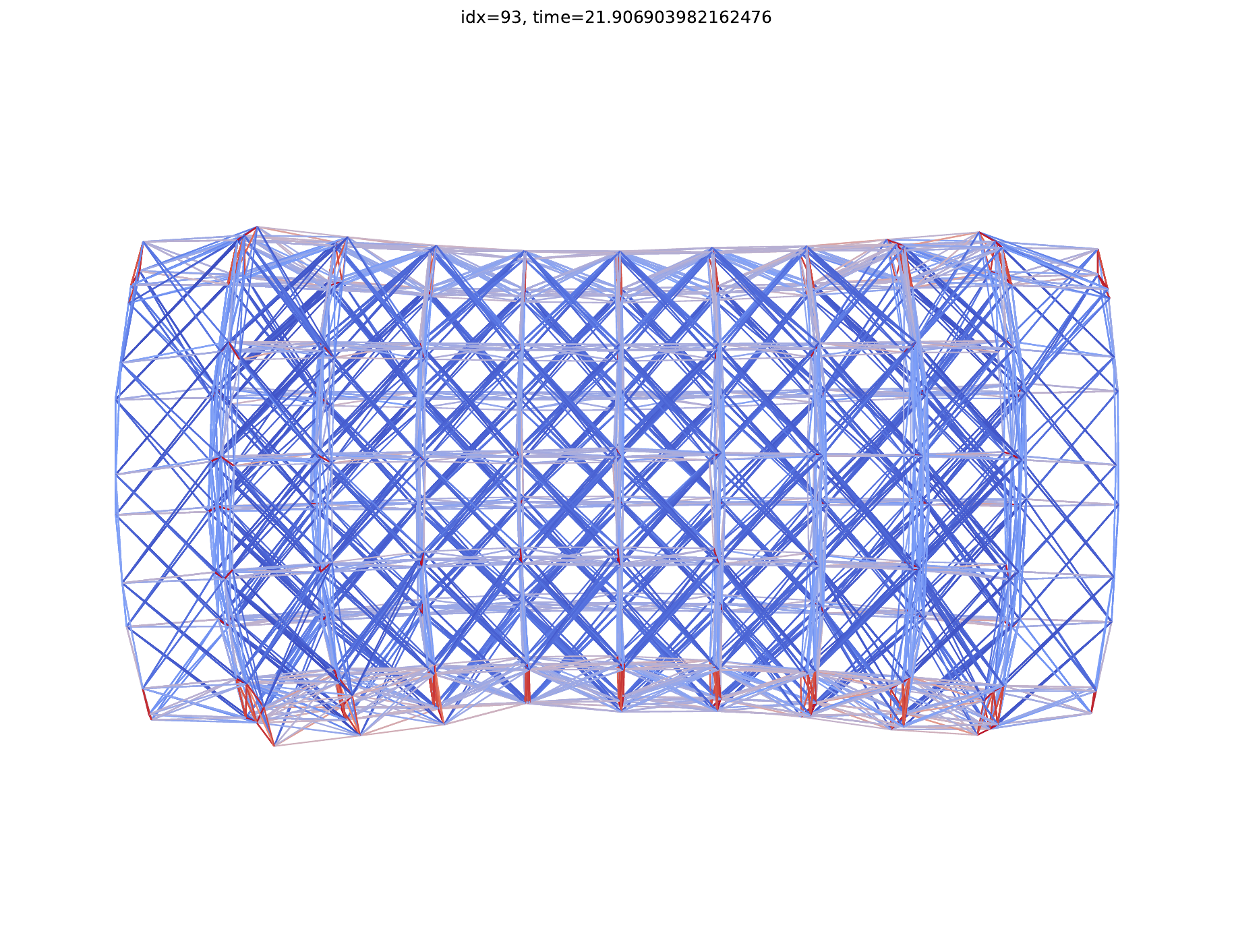} &
\imgcell{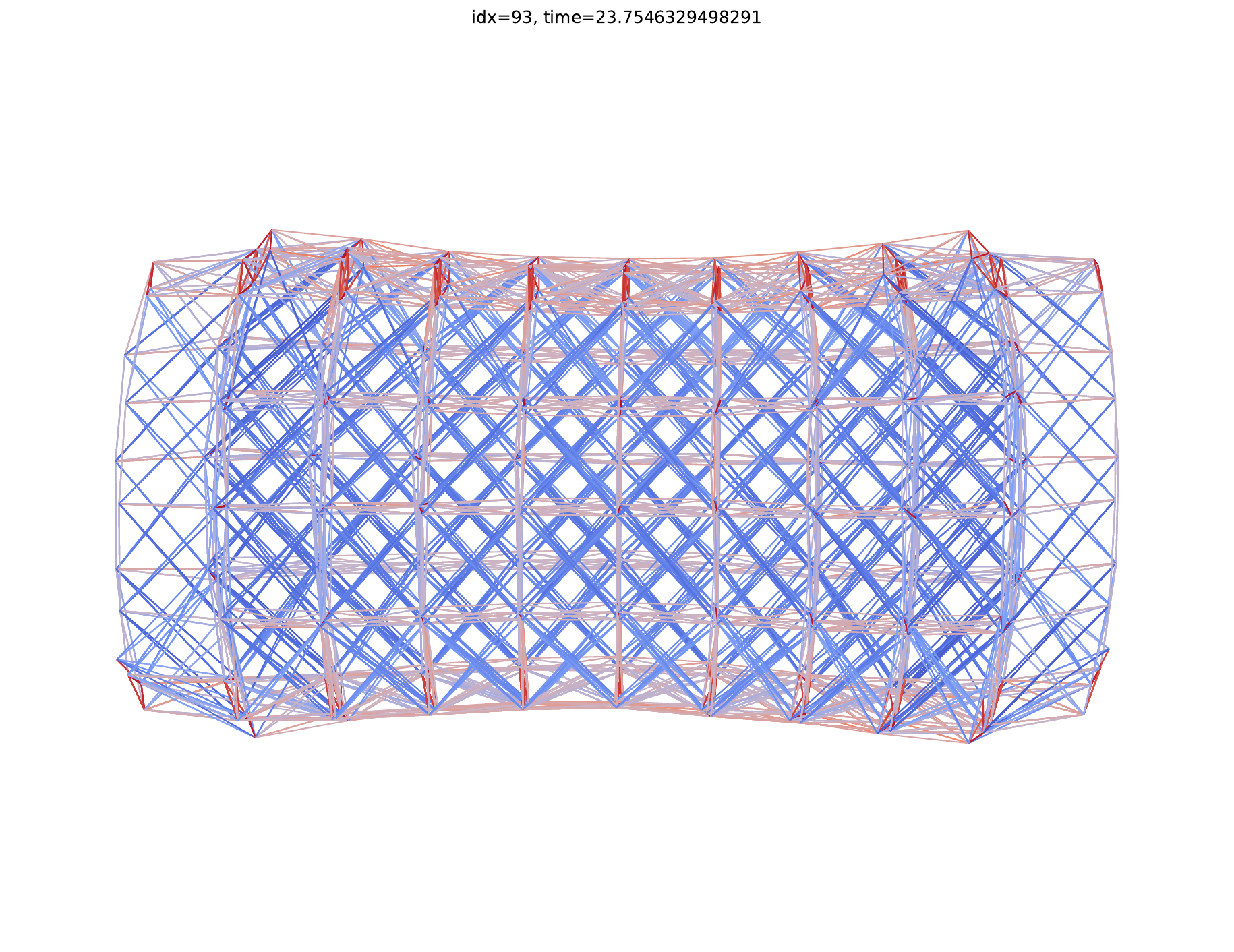} &
\imgcell{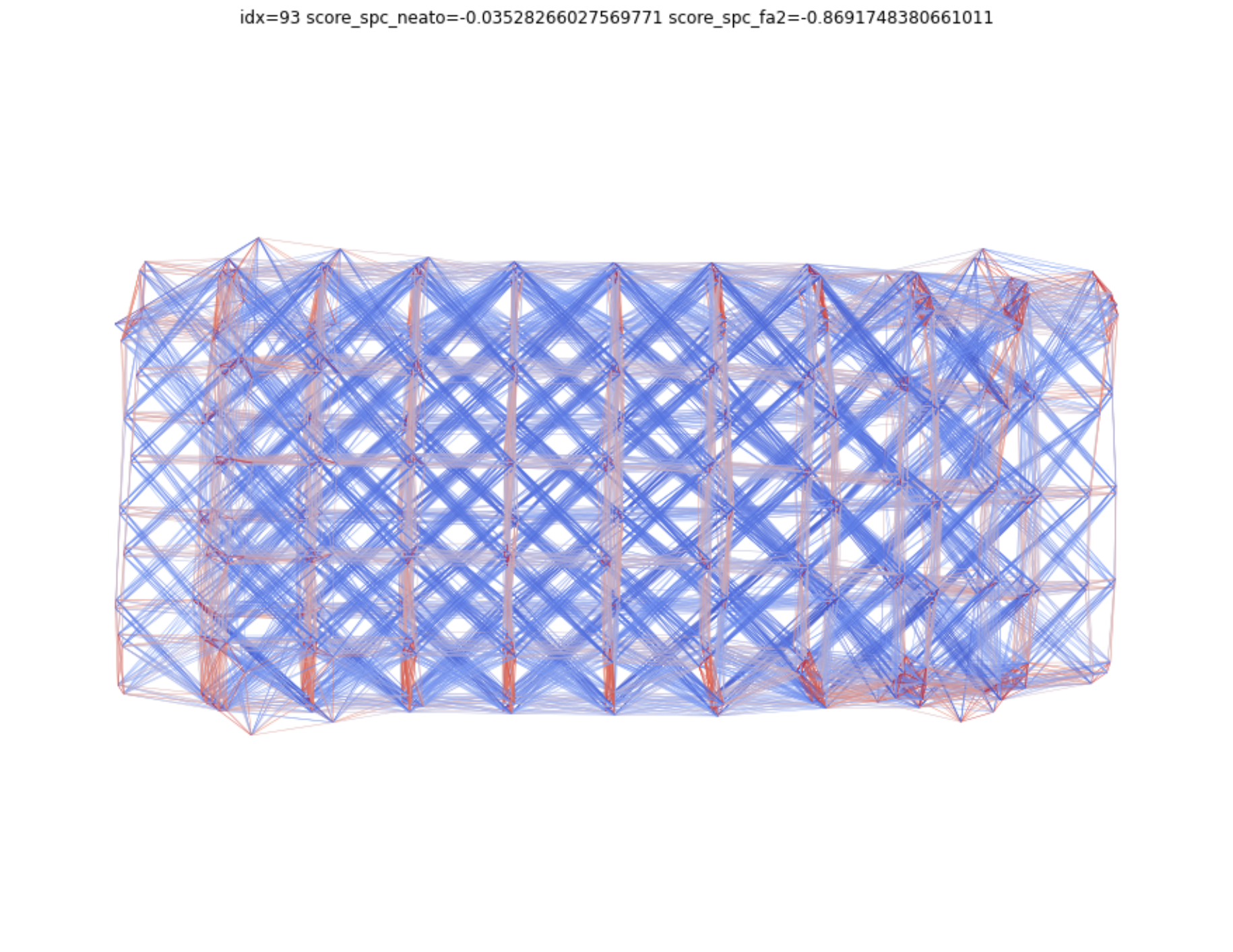} \\

&
t = 0.17s &
t = 49.33s &
t = 30.13s &
t = 1.45s &
t = 7200.00s &
t = 1.08s &
t = 1.53s &
t = 1.78s &
t = 1.54s &
t = 1.91s &
t = 1.75s &
t = 1.17s \\

\makecell{\bfseries nnc666\\N = 666\\M = 2148} &
\imgcell{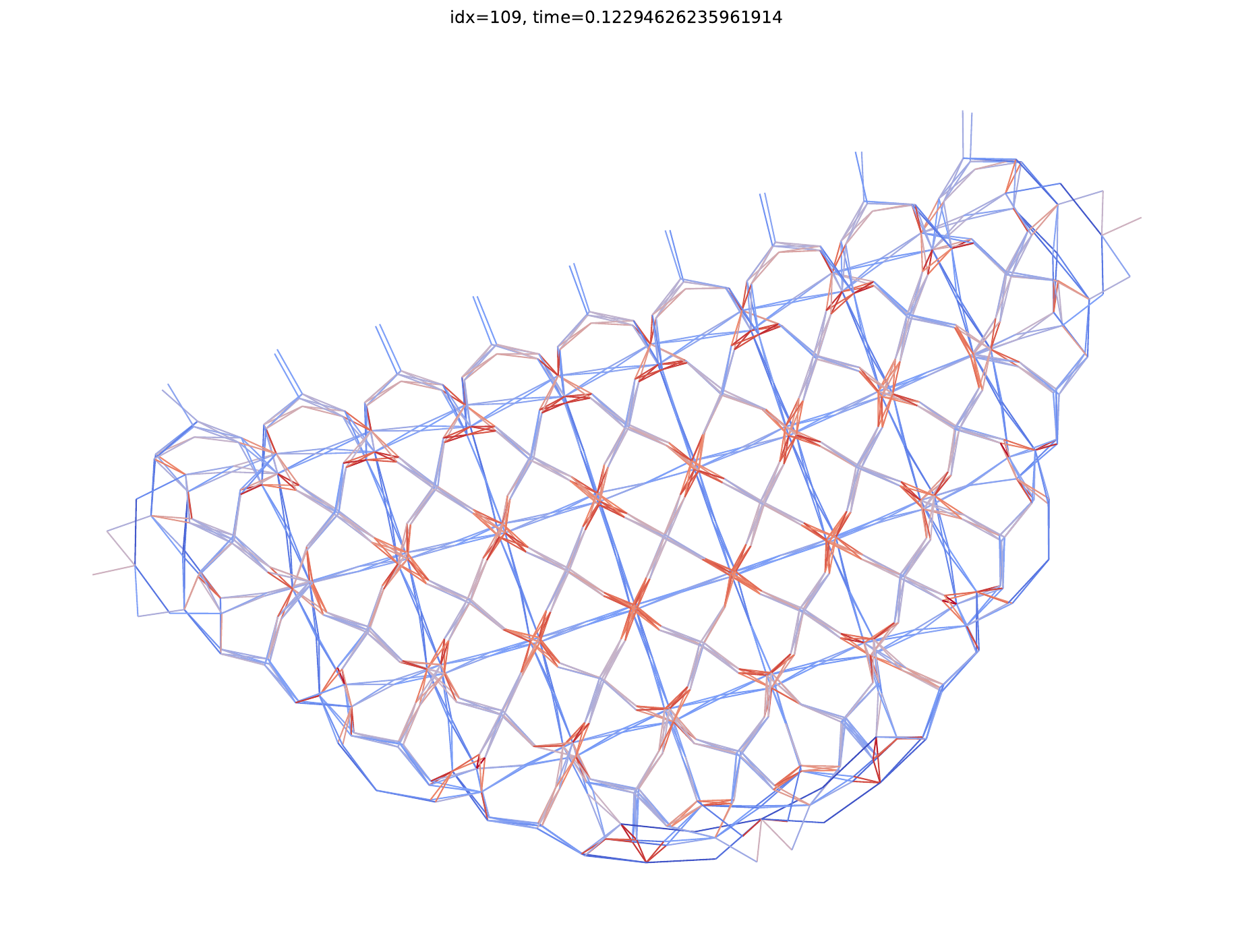} &
\imgcell{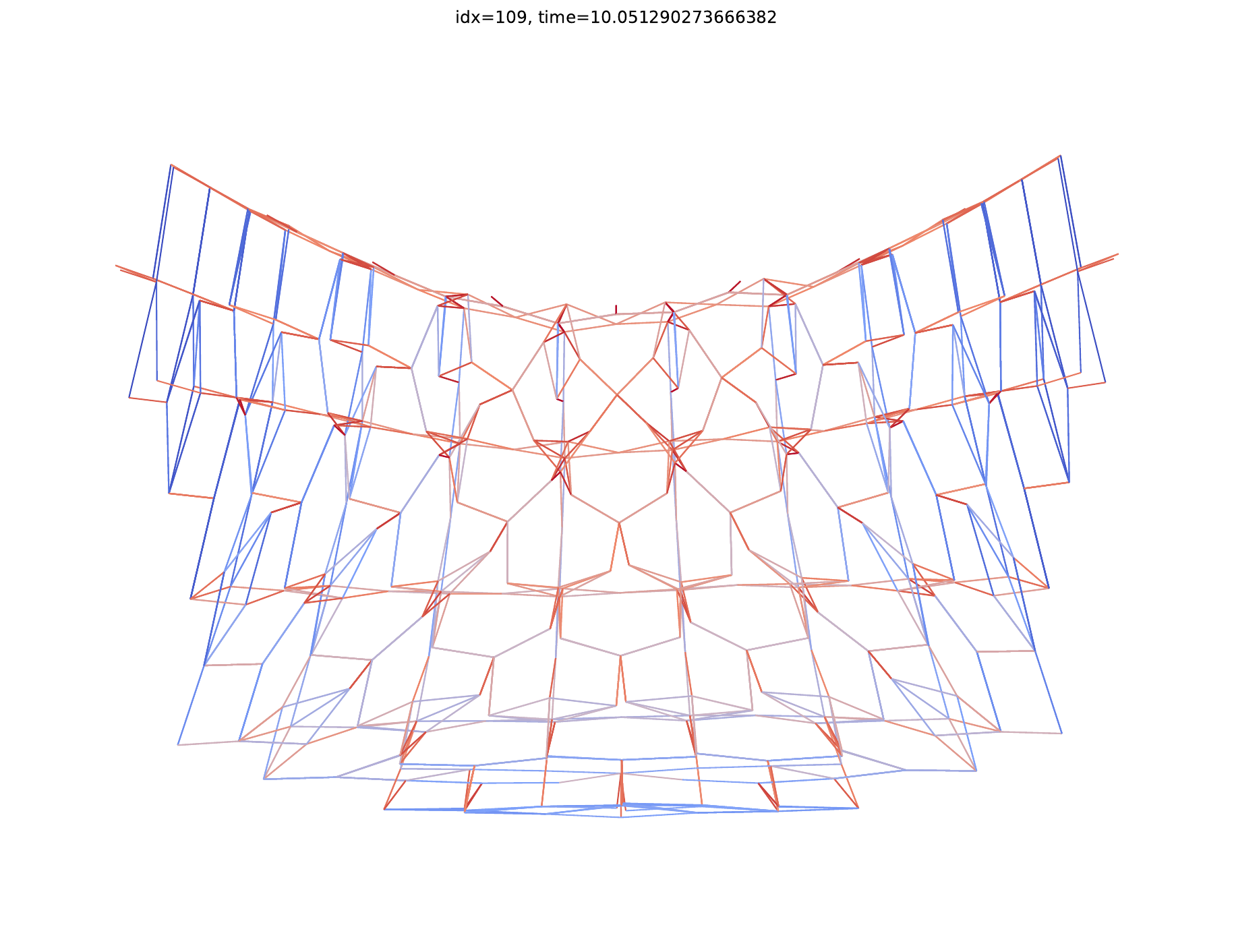} &
\imgcell{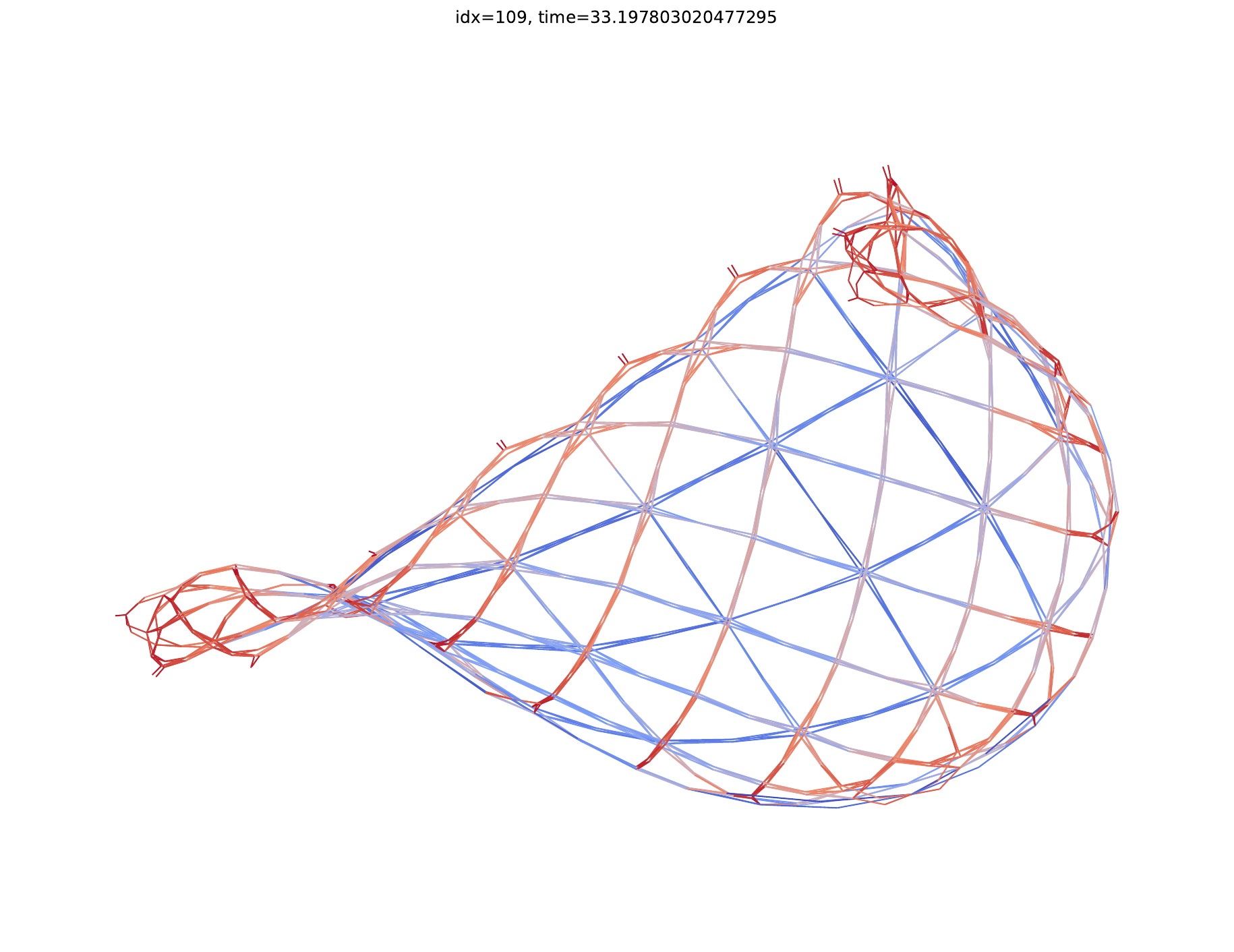} &
\imgcell{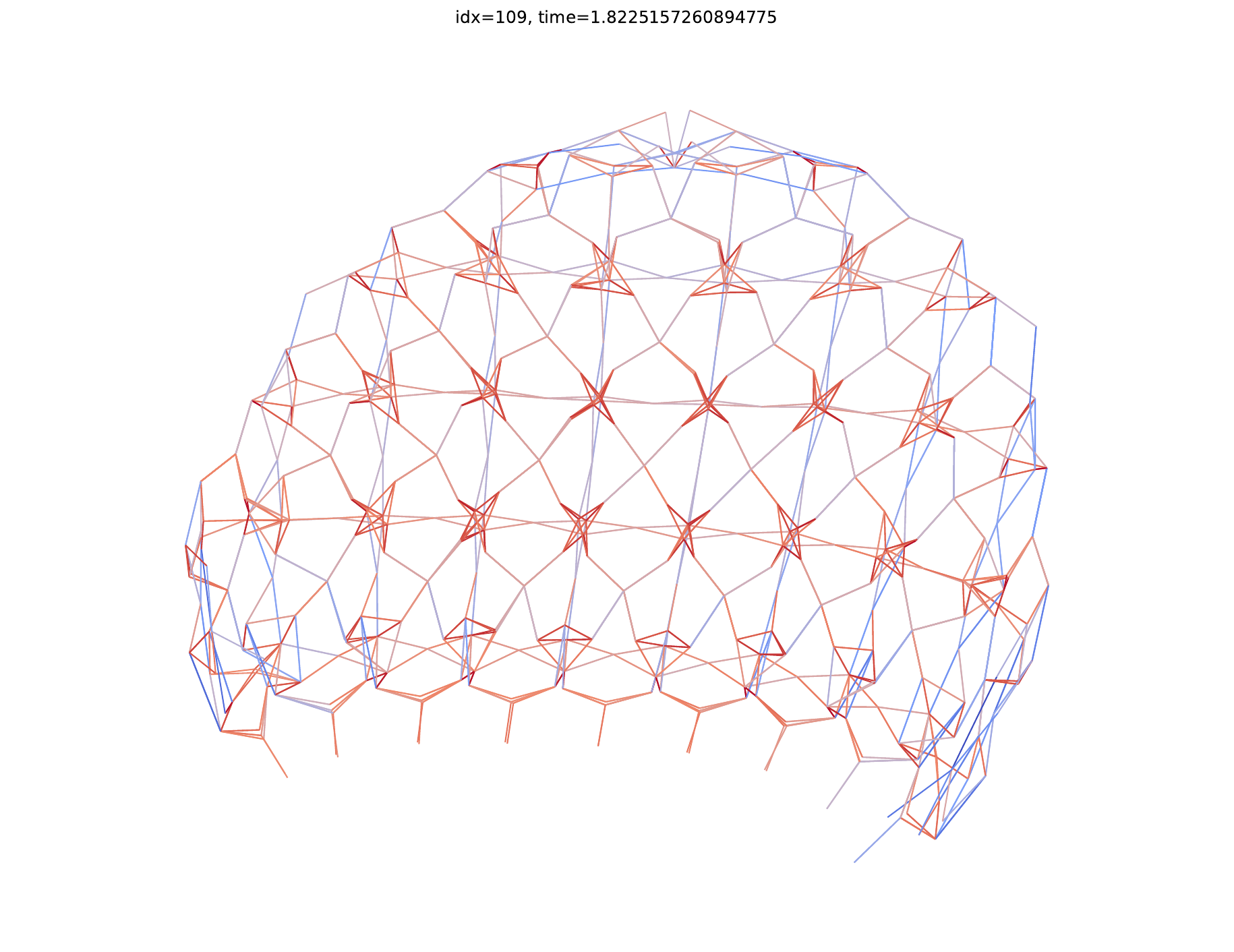} &
\imgcell{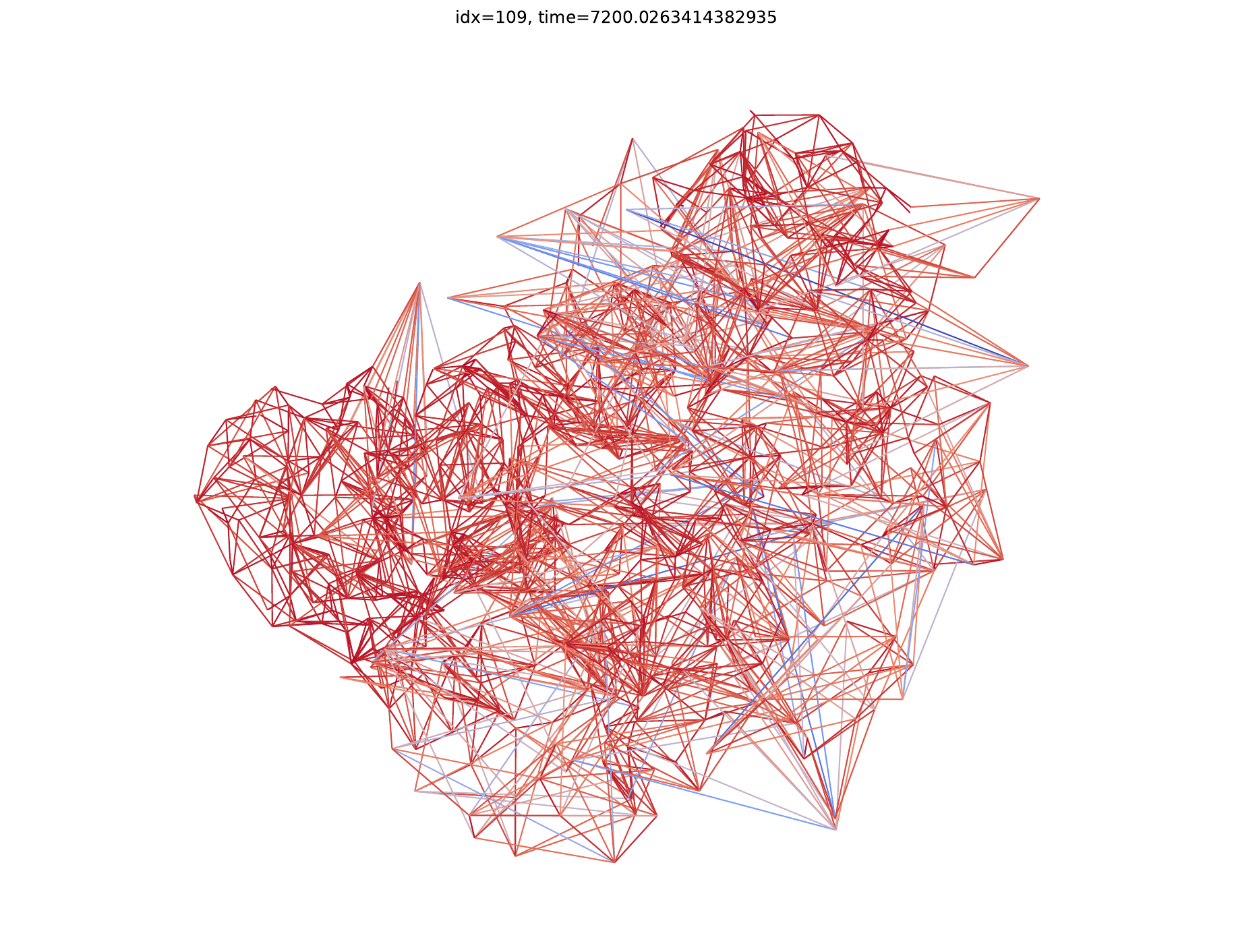} &
\imgcell{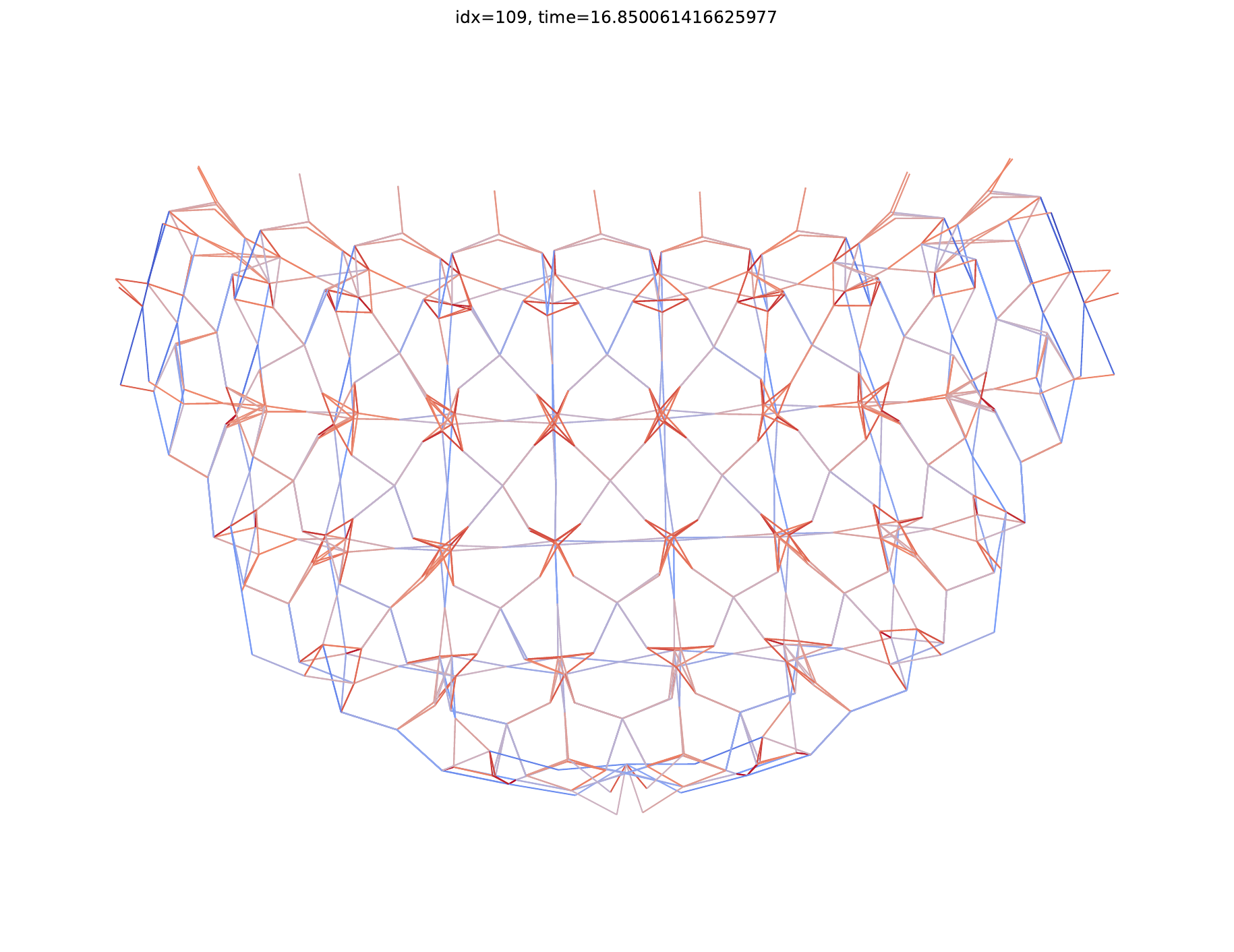} &
\imgcell{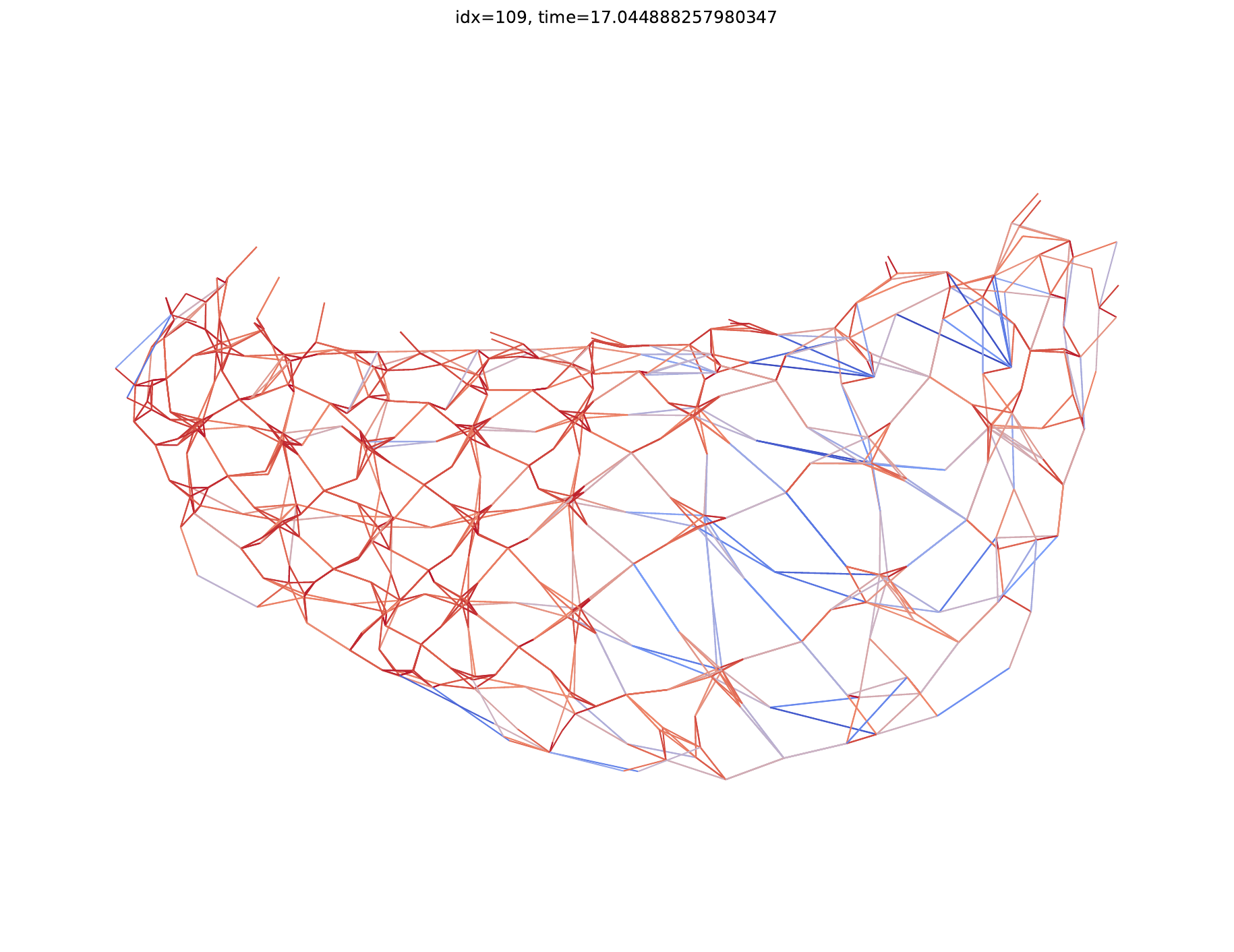} &
\imgcell{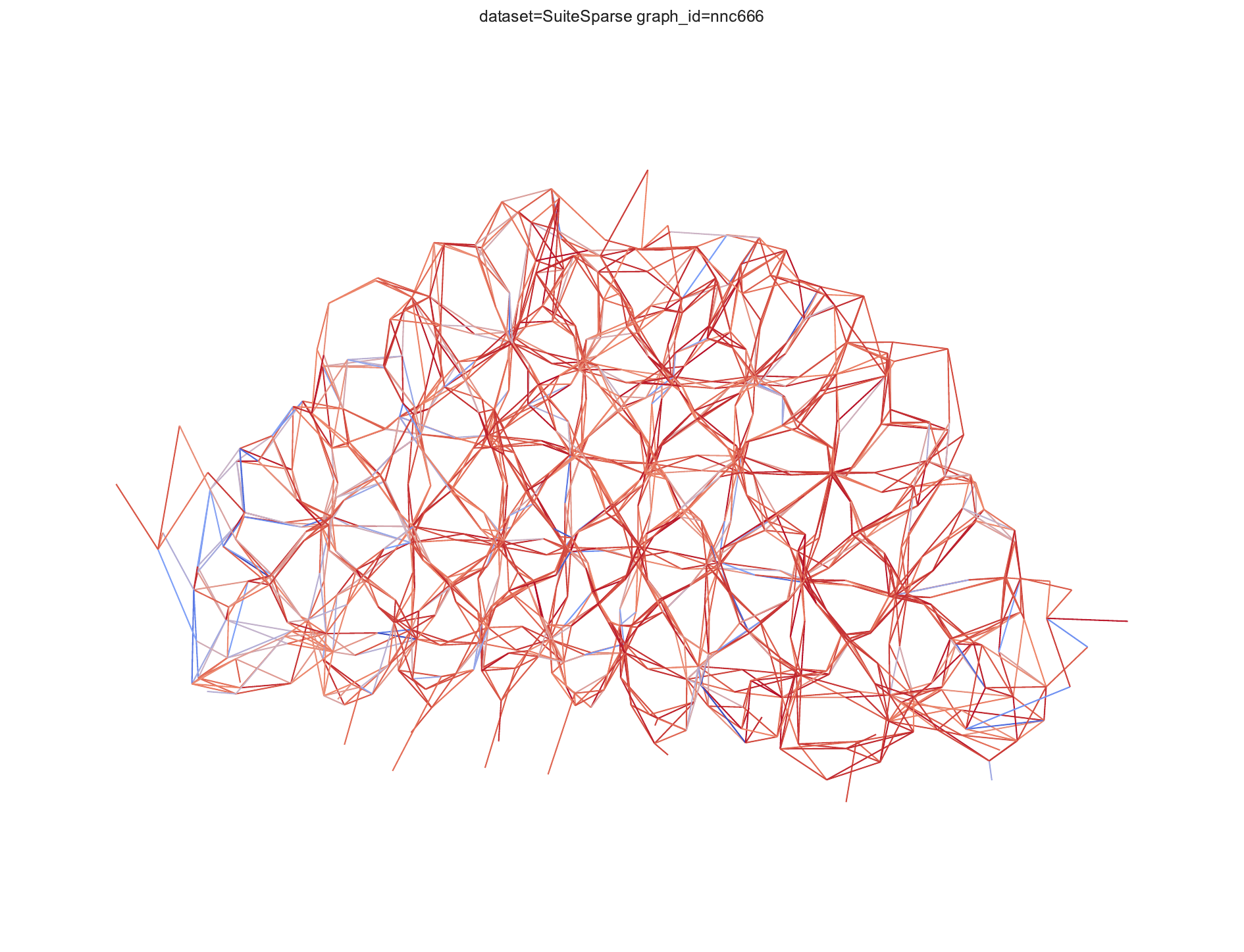} &
\imgcell{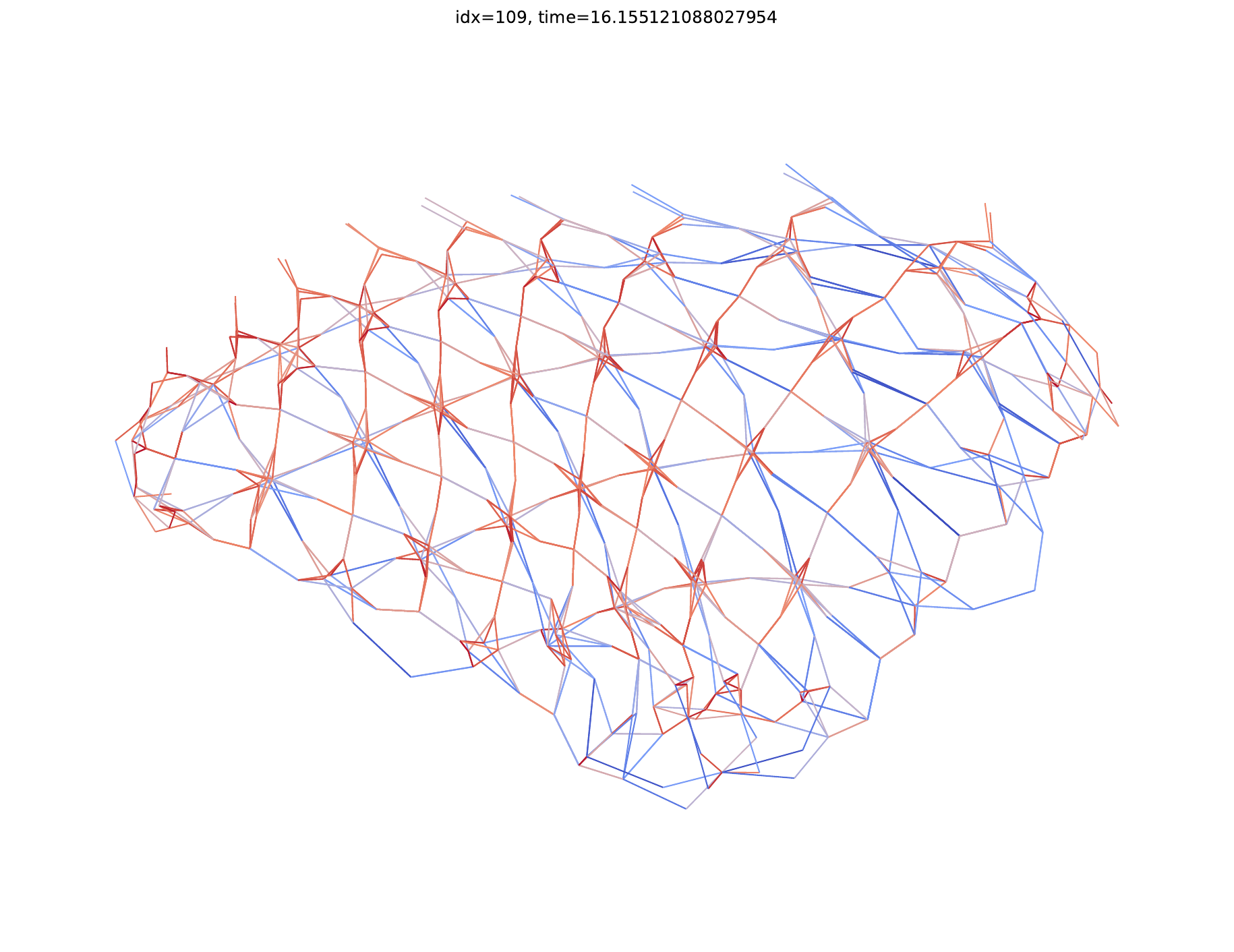} &
\imgcell{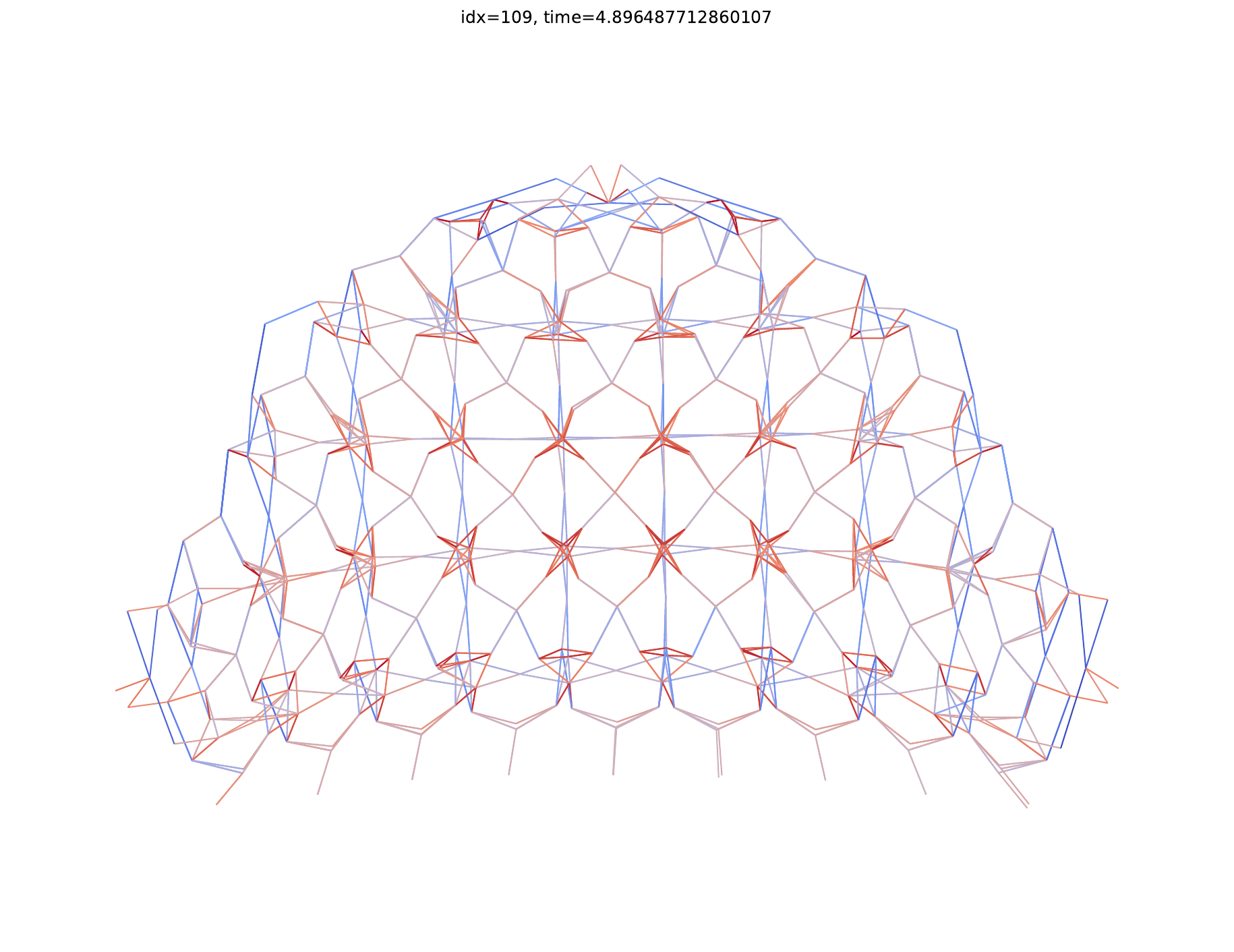} &
\imgcell{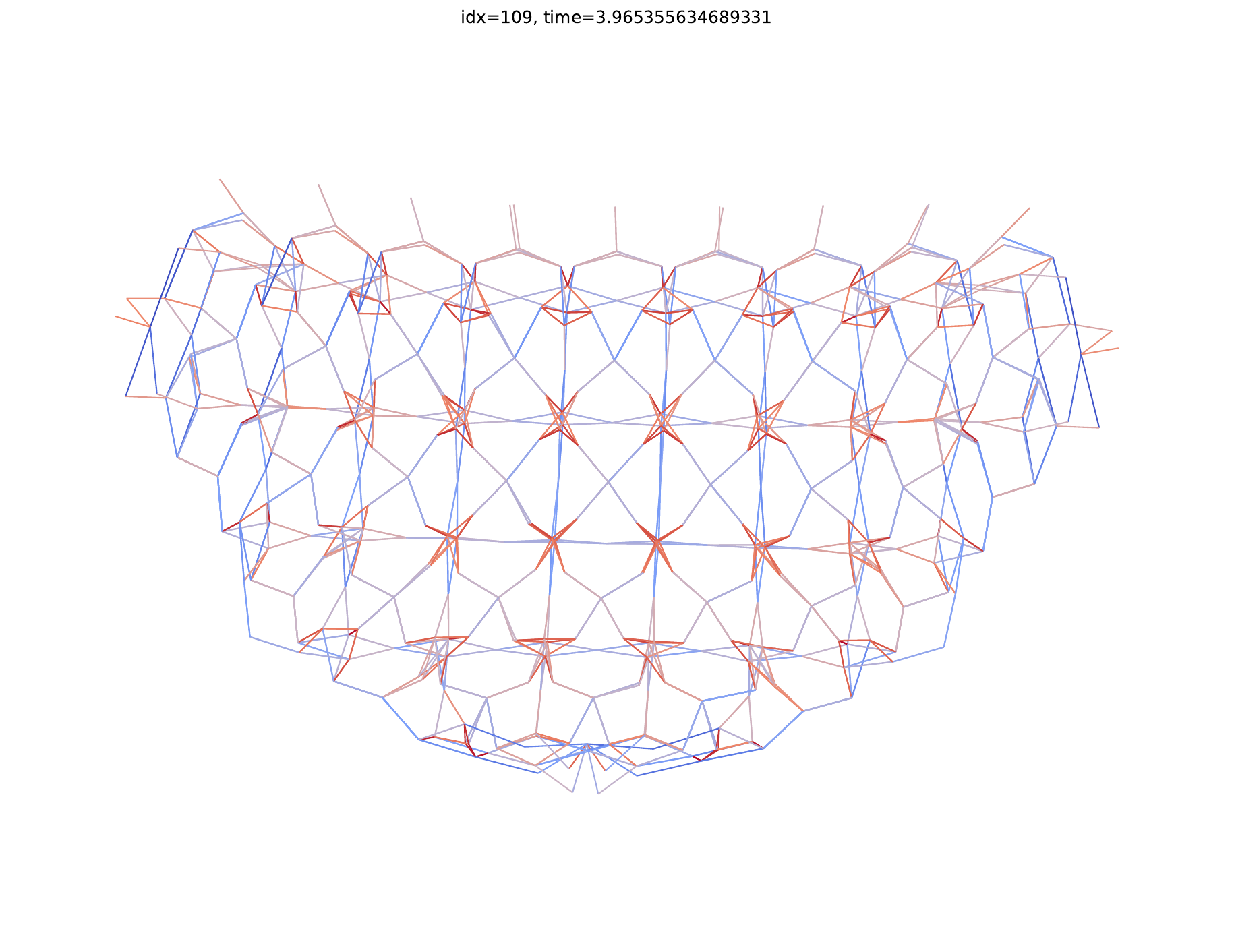} &
\imgcell{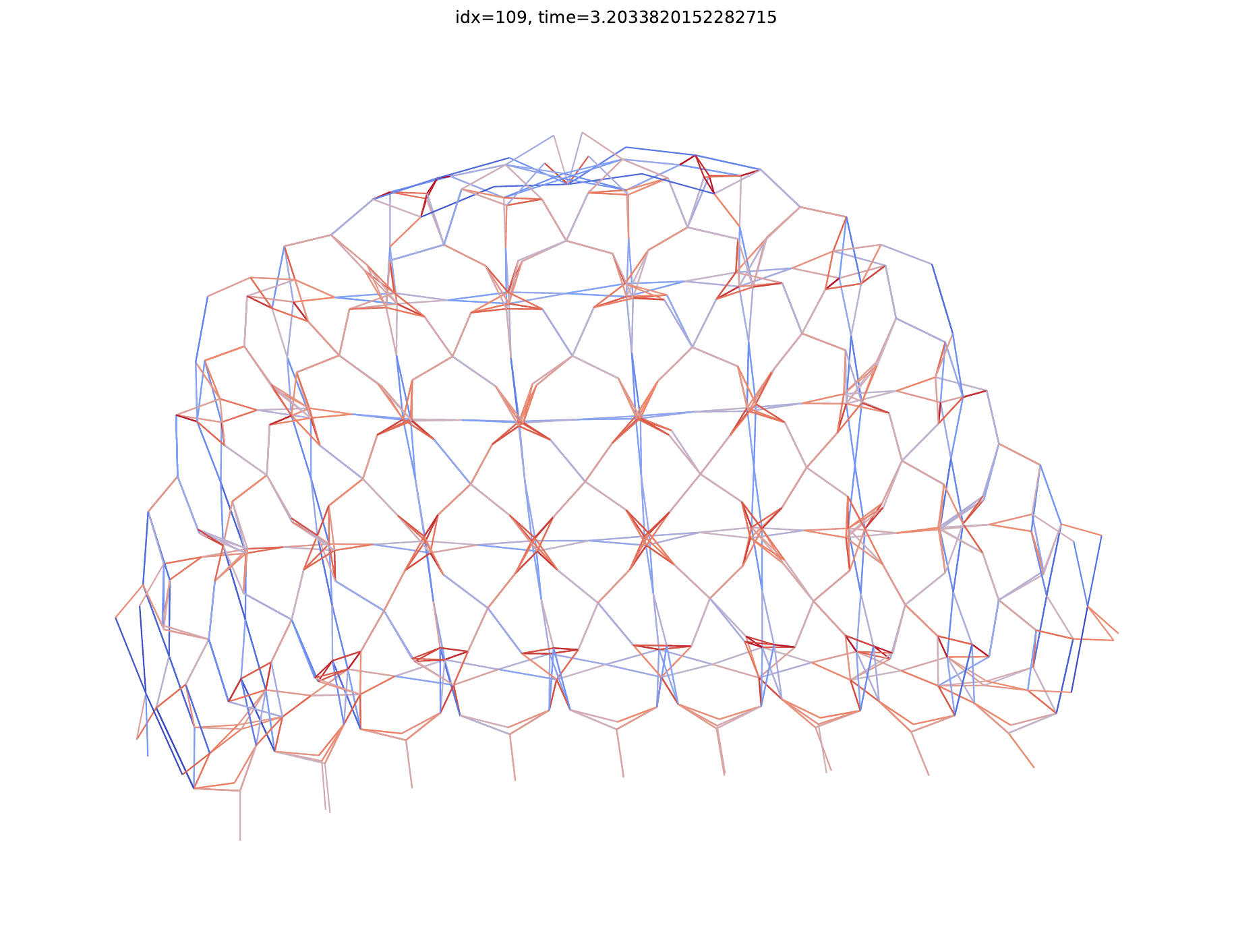} \\

&
t = 0.12s &
t = 10.05s &
t = 33.20s &
t = 1.82s &
t = 7200.00s &
t = 1.72s &
t = 1.90s &
t = 1.88s &
t = 1.60s &
t = 1.58s &
t = 1.62s &
t = 1.54s \\

\makecell{\bfseries msc00726\\N = 726\\M = 16896} &
\imgcell{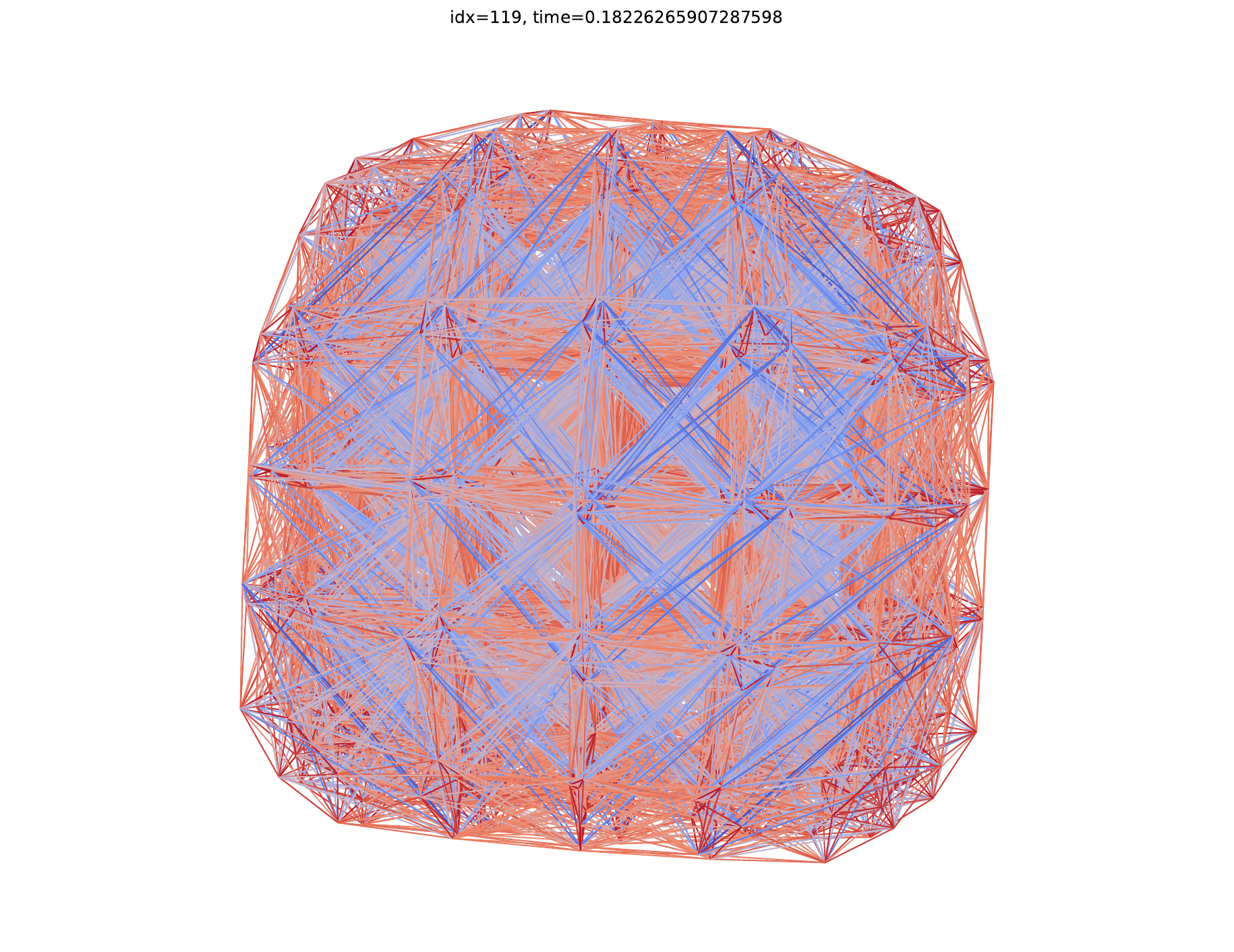} &
\imgcell{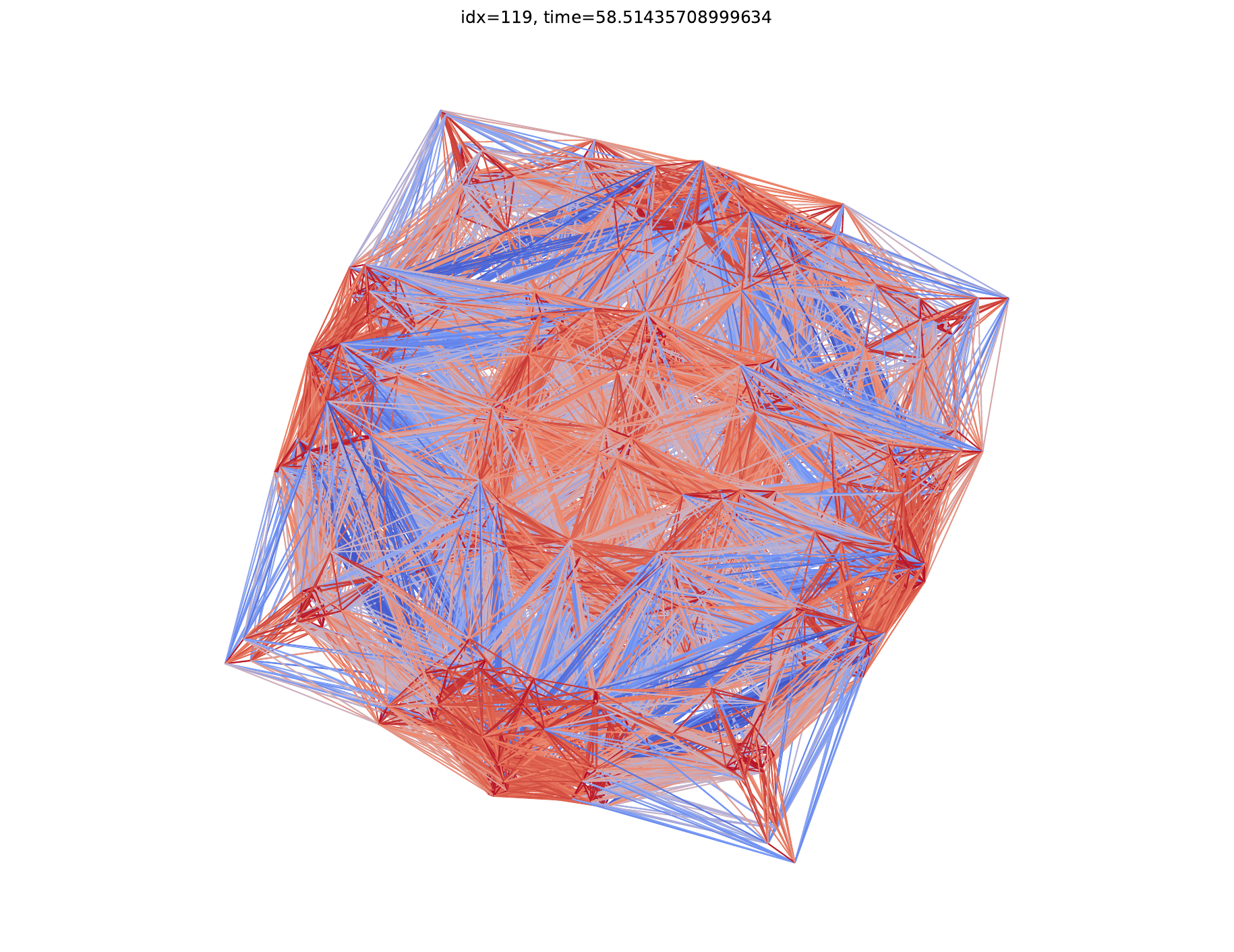} &
\imgcell{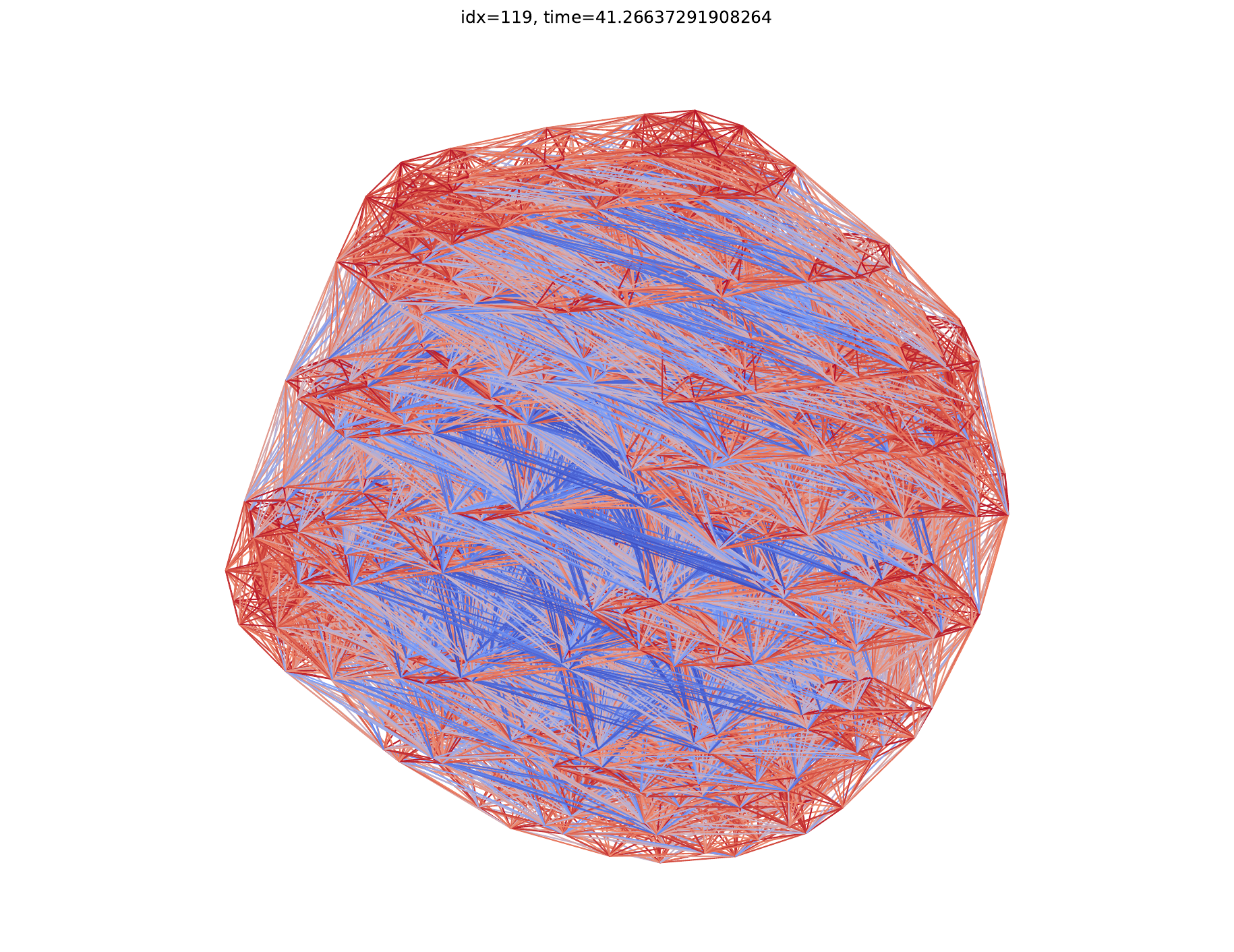} &
\imgcell{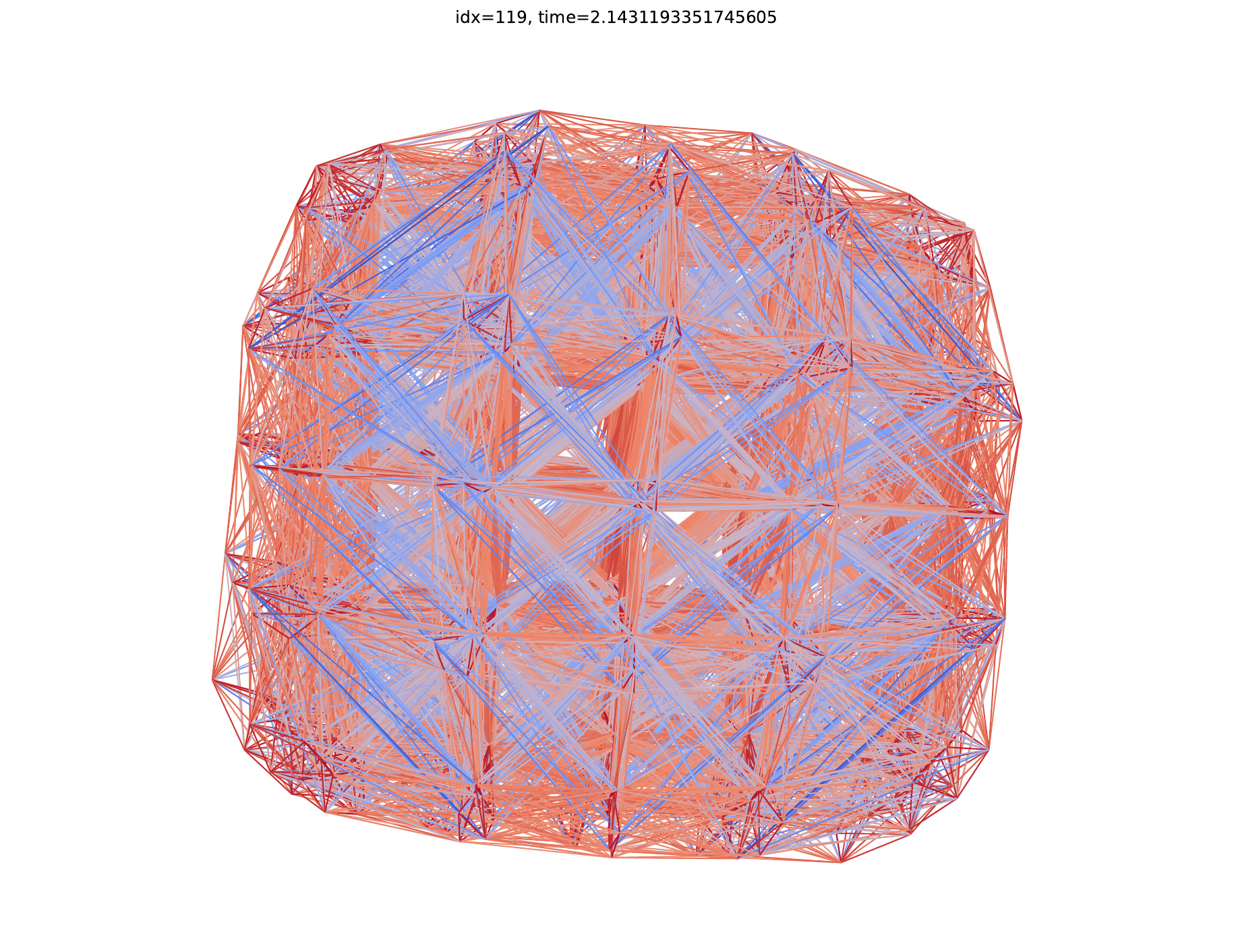} &
\imgcell{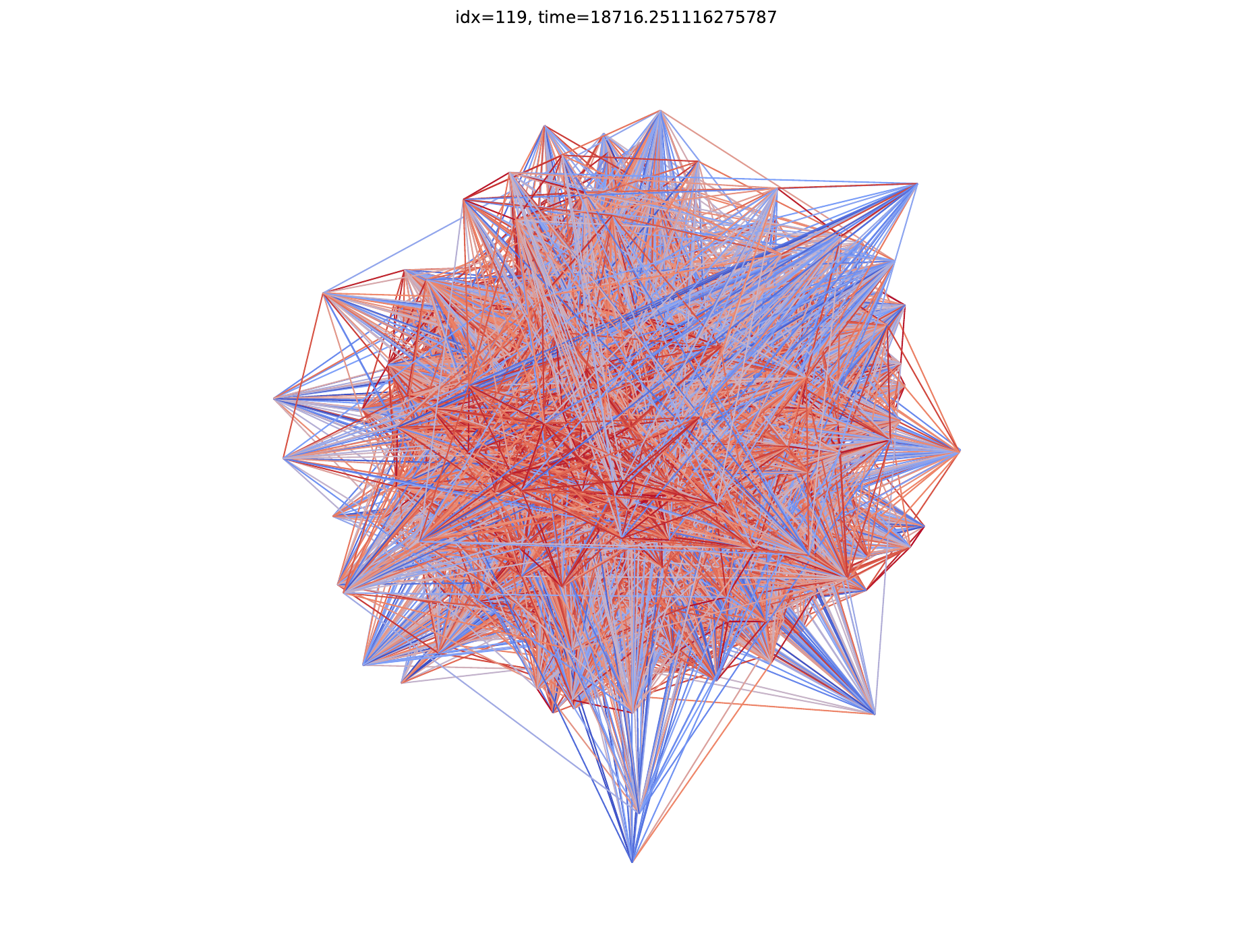} &
\imgcell{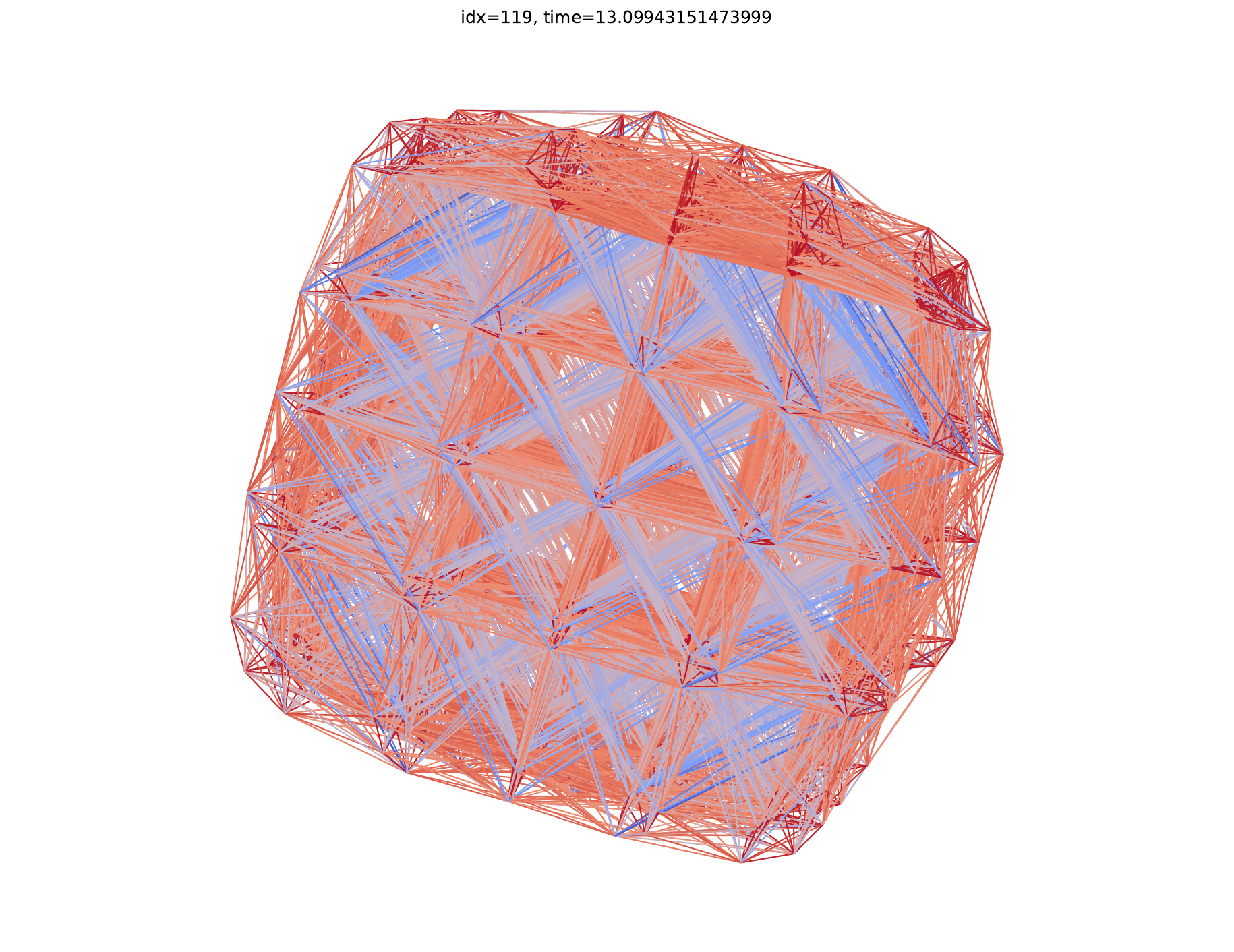} &
\imgcell{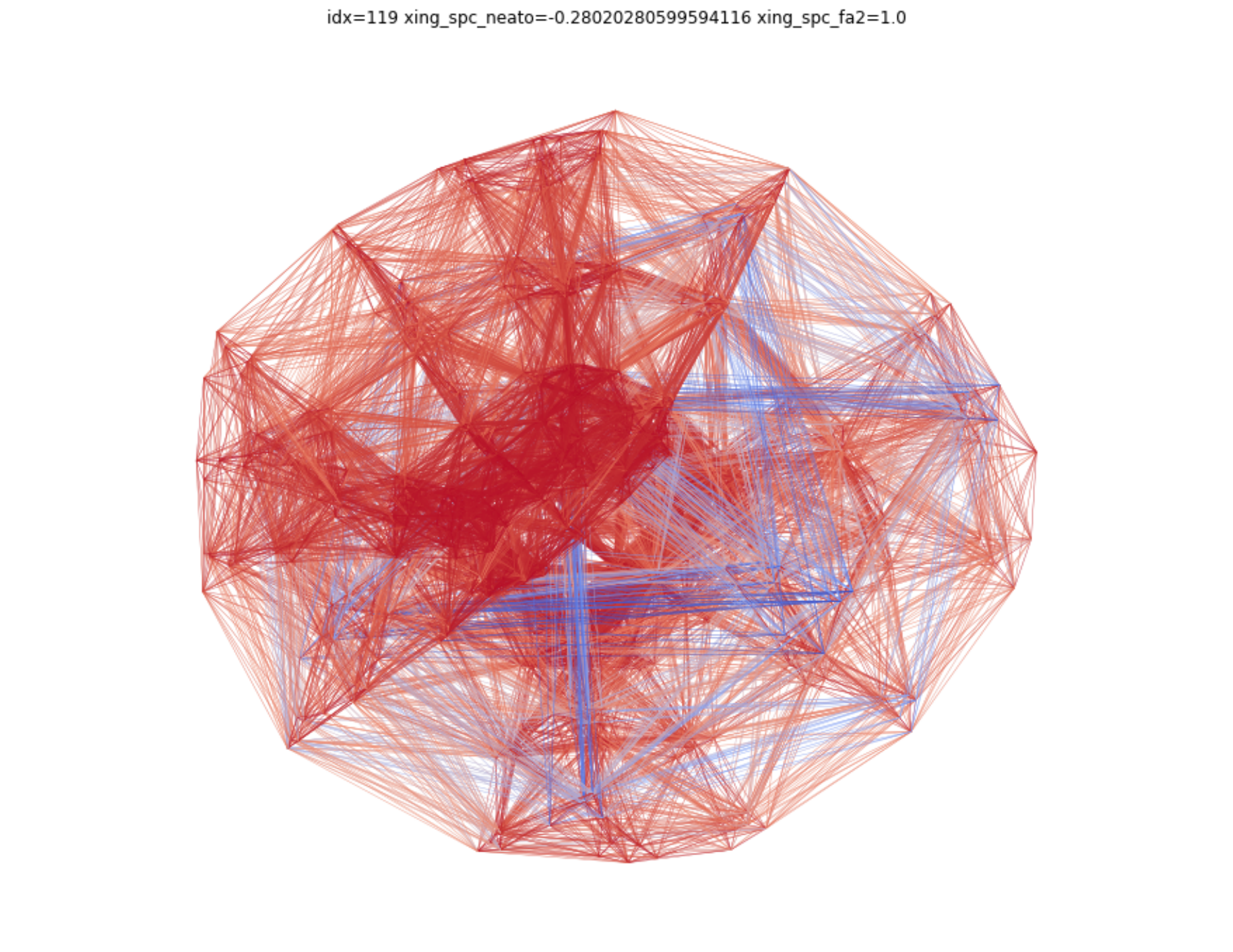} &
\imgcell{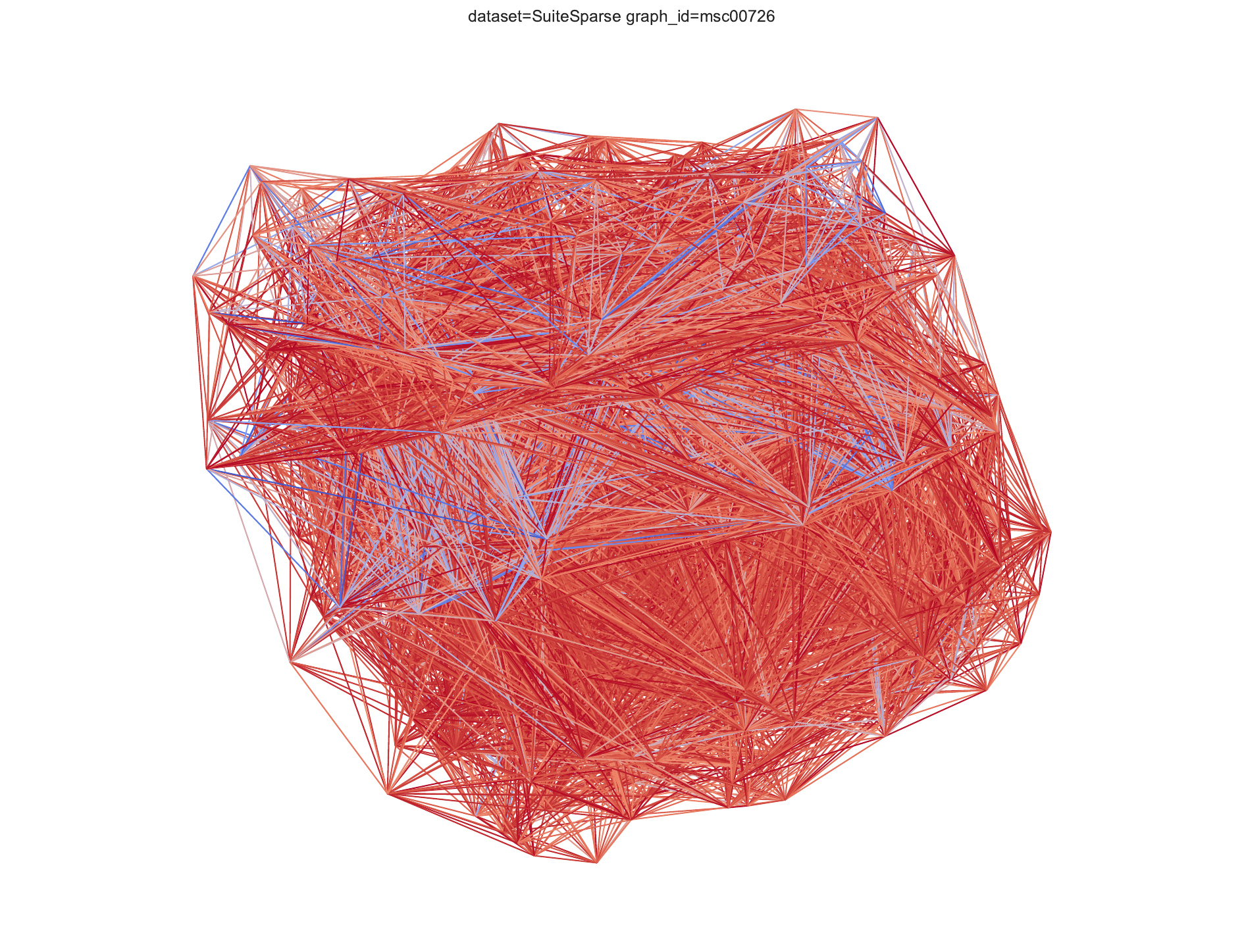} &
\imgcell{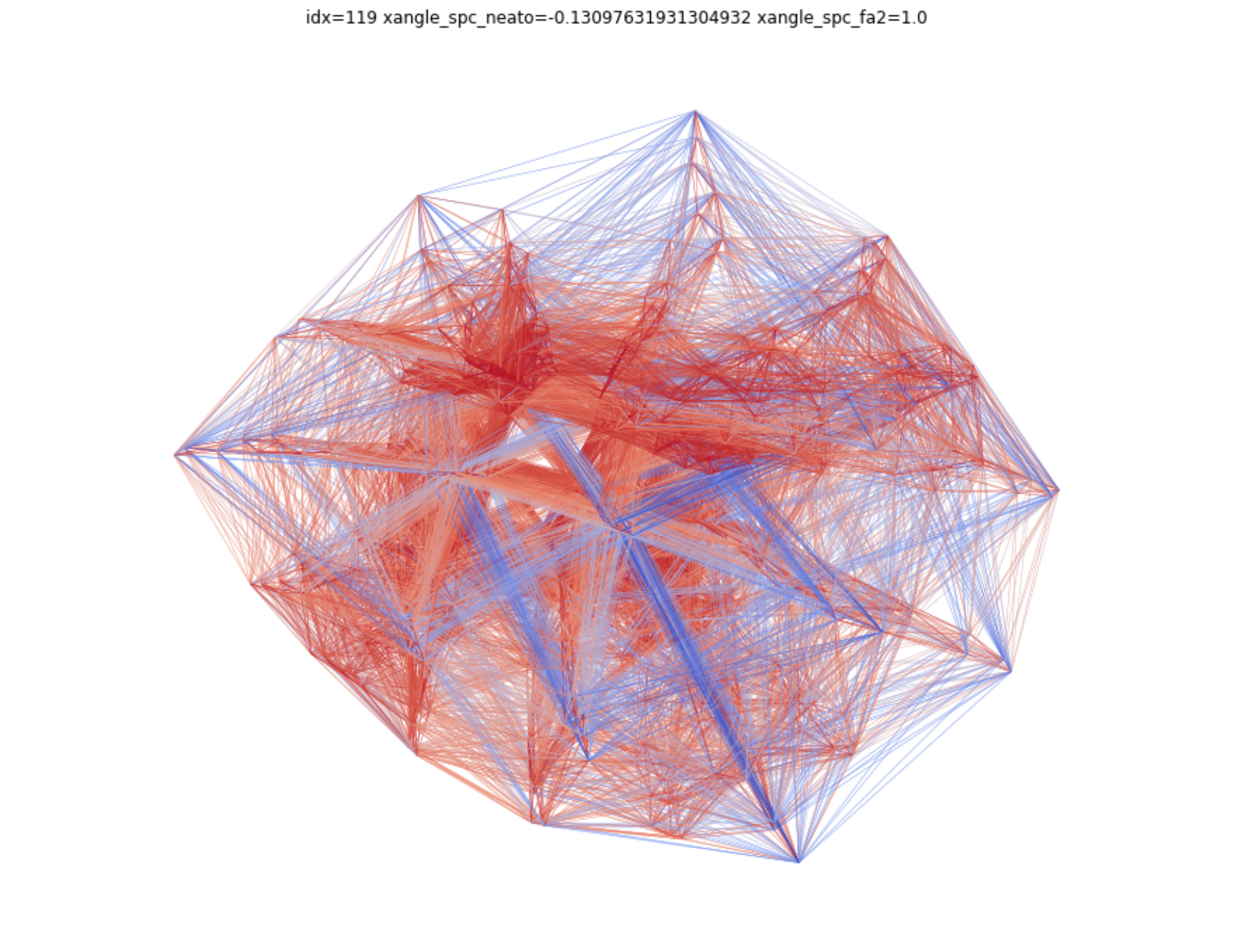} &
\imgcell{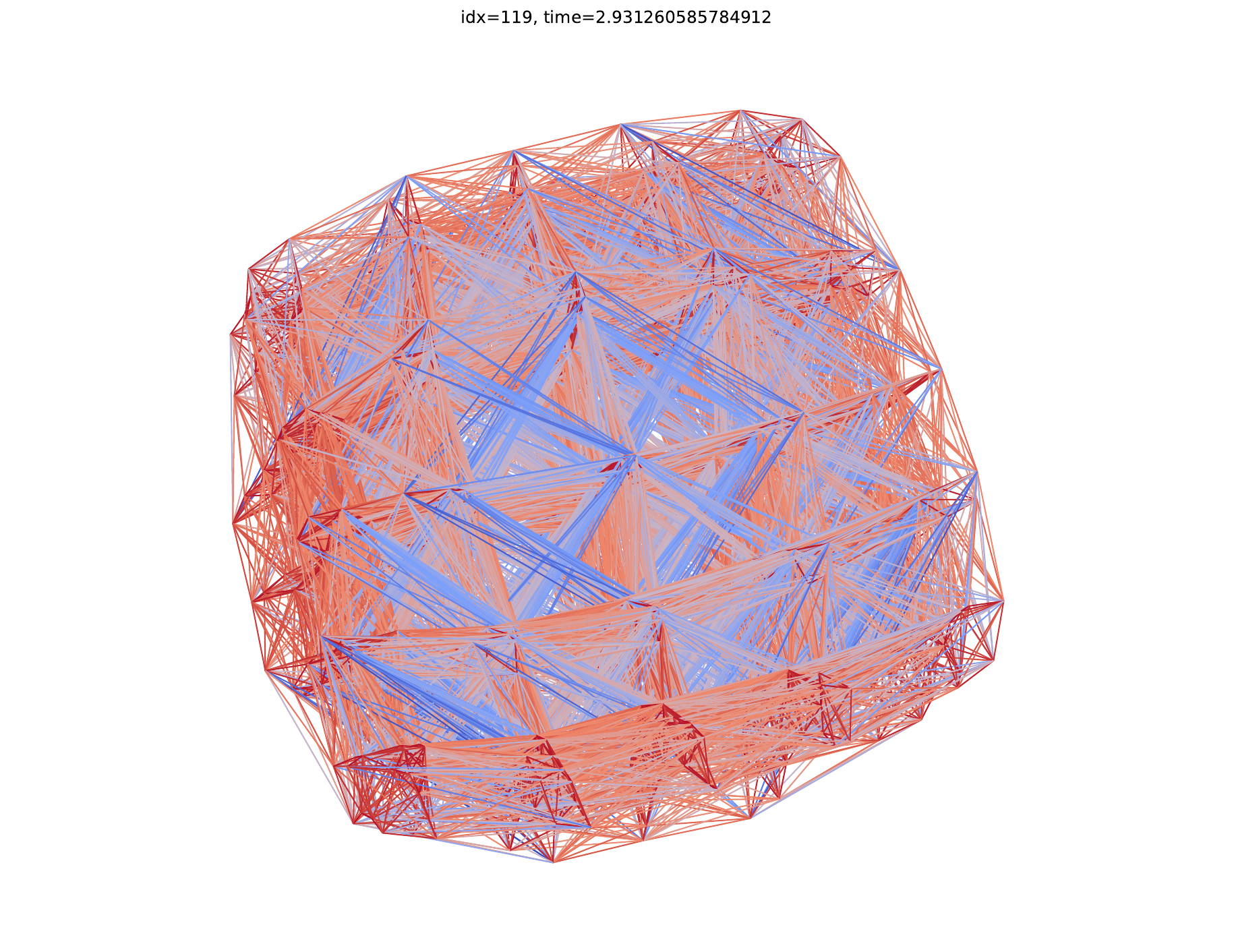} &
\imgcell{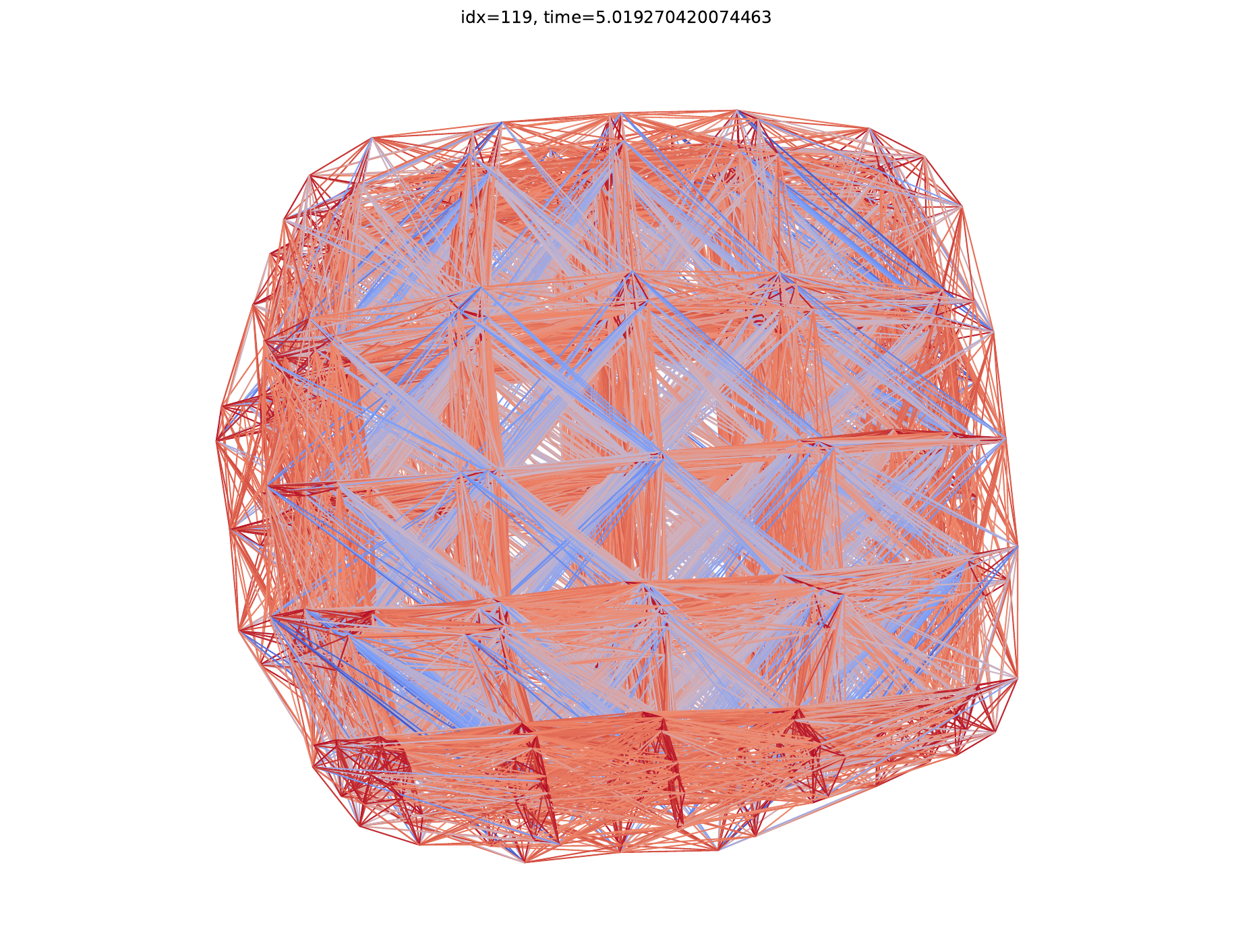} &
\imgcell{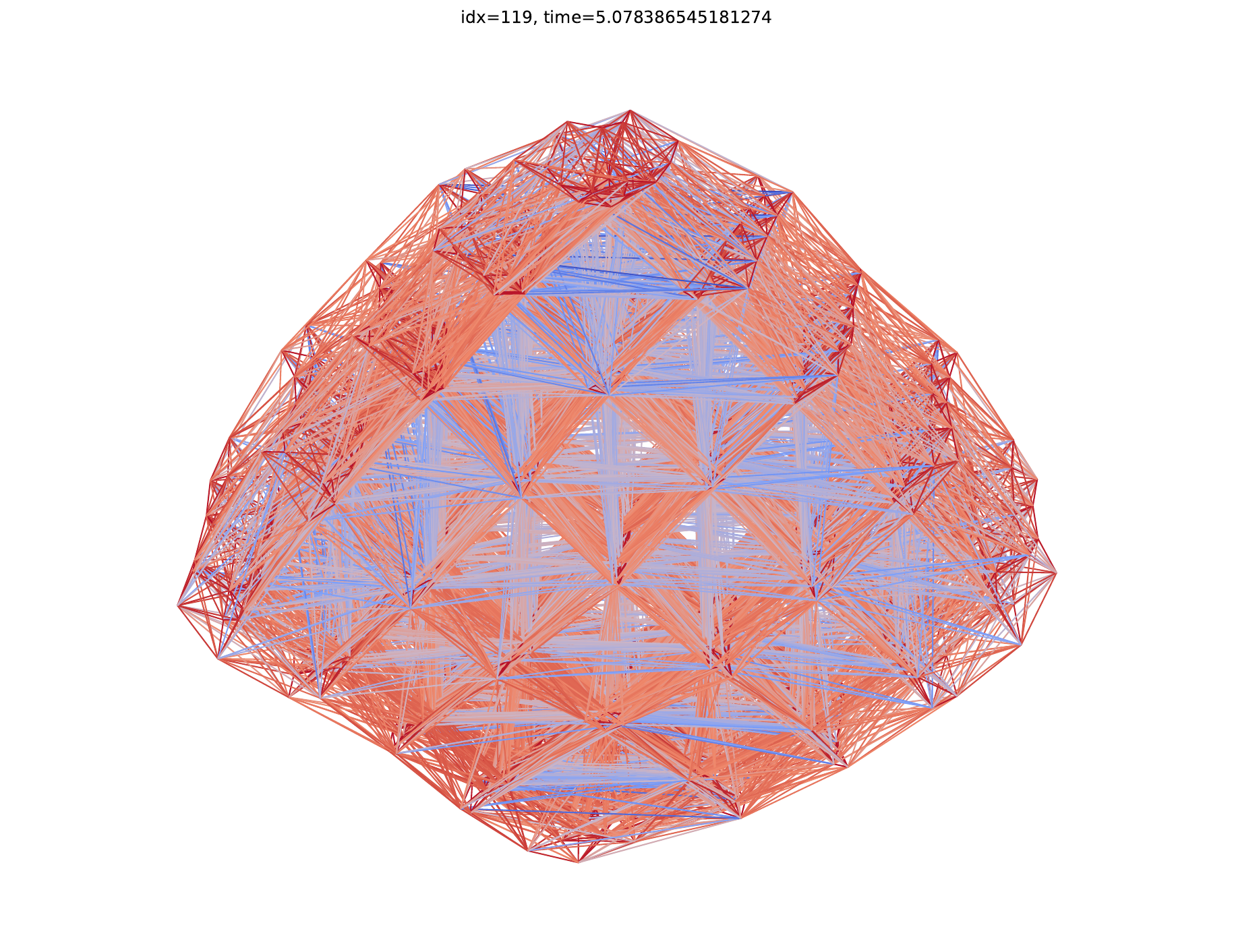} \\

&
t = 0.18s &
t = 58.51s &
t = 41.27s &
t = 2.14s &
t = 7200.00s &
t = 1.99s &
t = 2.16s &
t = 1.97s &
t = 1.64s &
t = 1.99s &
t = 2.31s &
t = 1.83s \\

\makecell{\bfseries dendrimer\\N = 730\\M = 31147} &
\imgcell{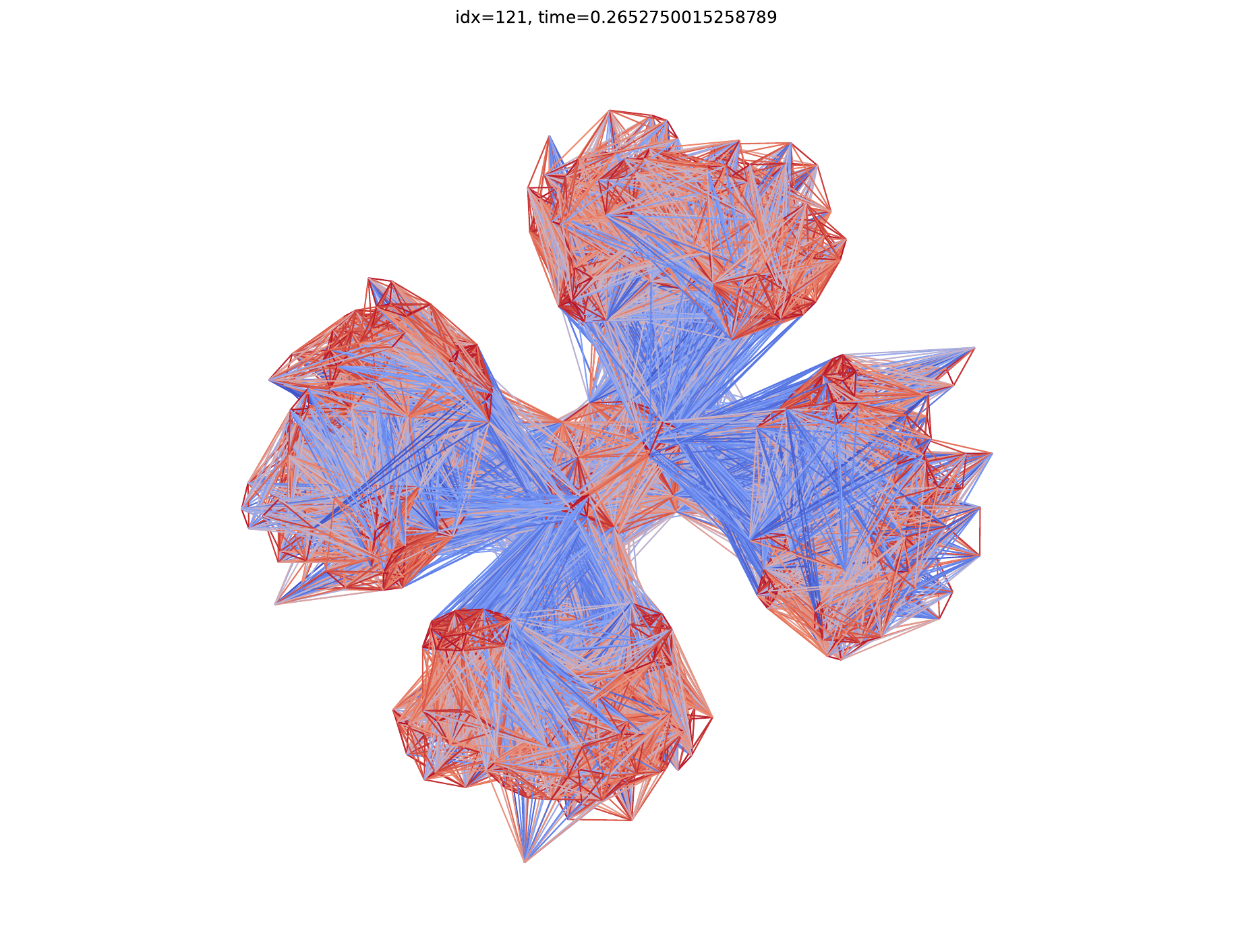} &
\imgcell{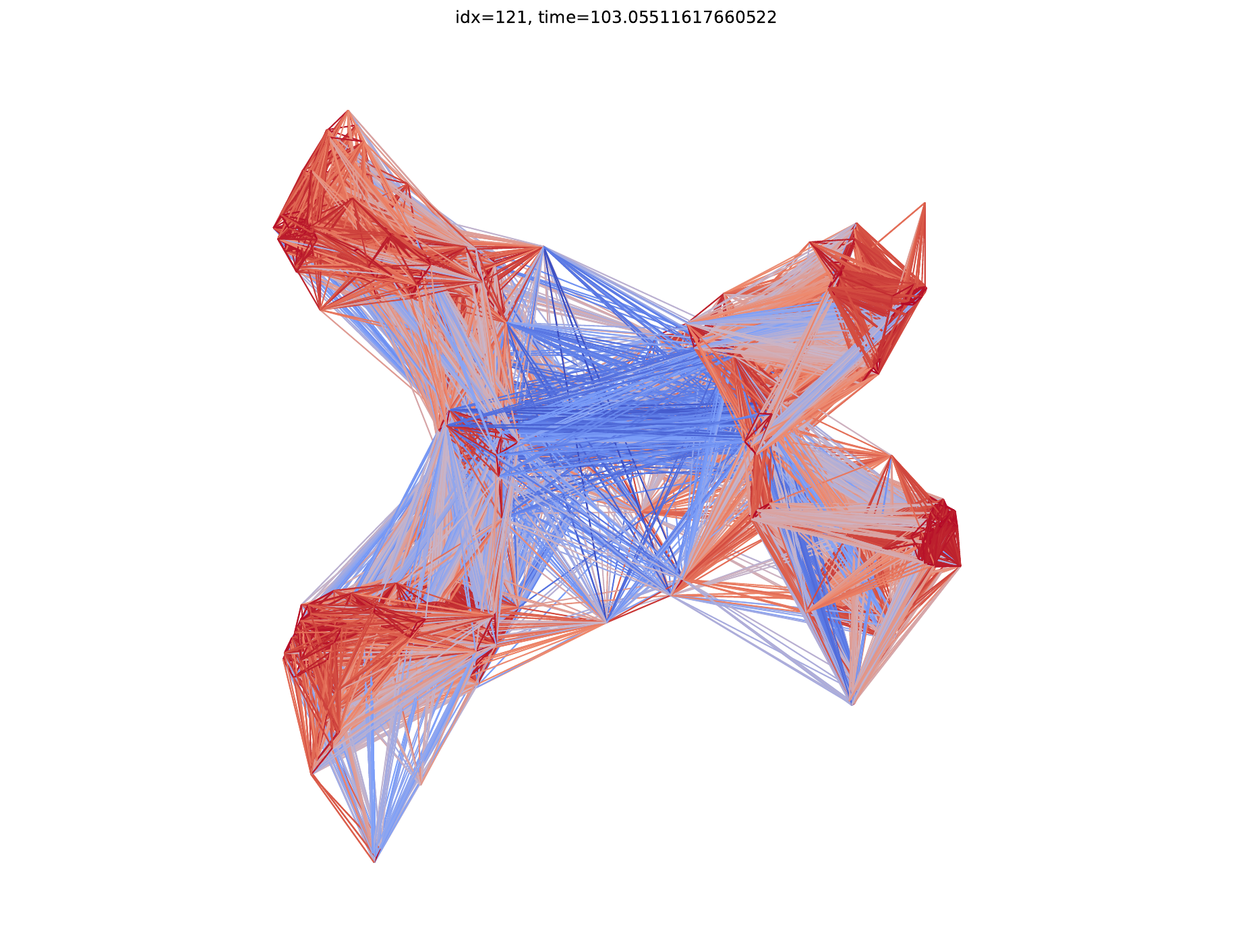} &
\imgcell{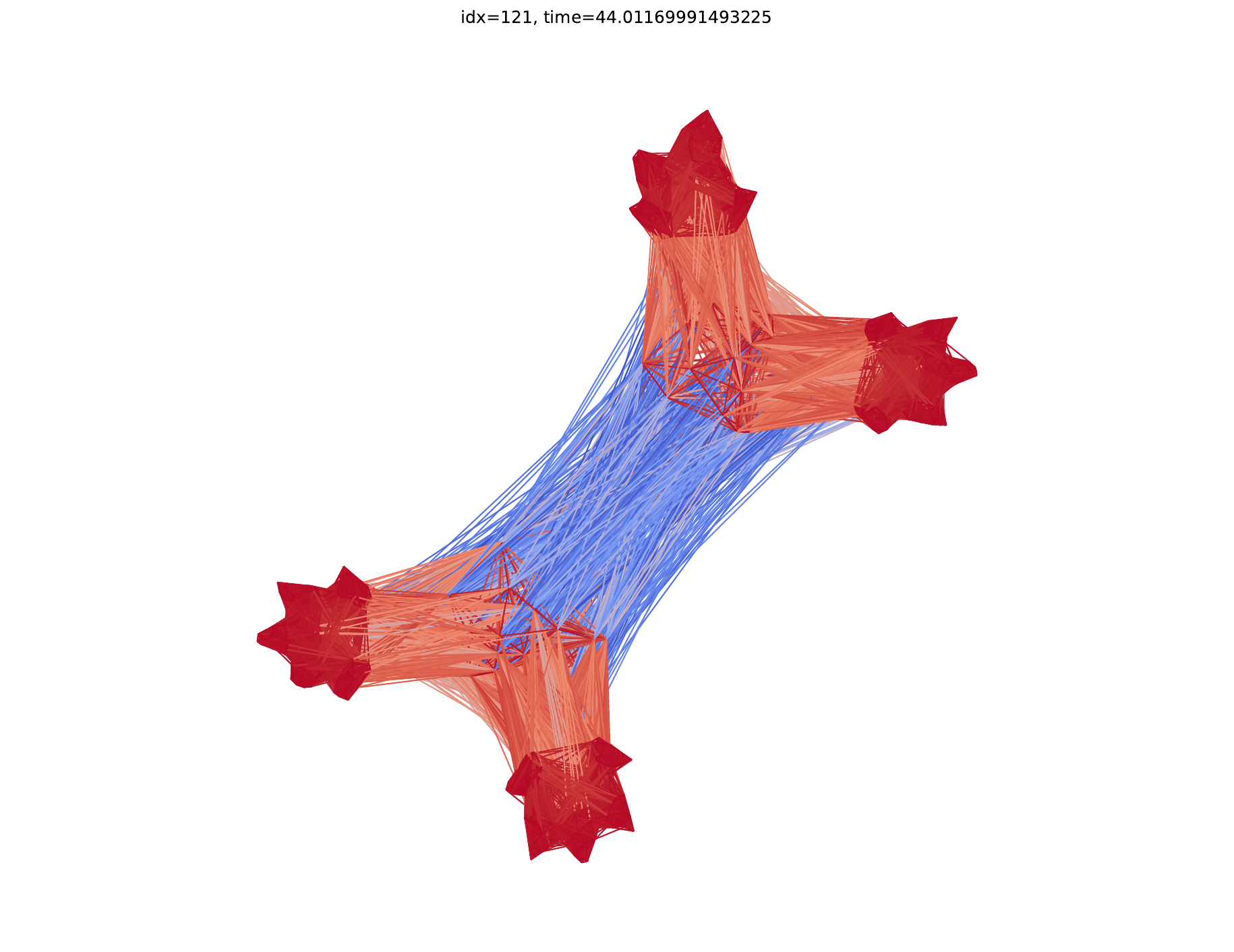} &
\imgcell{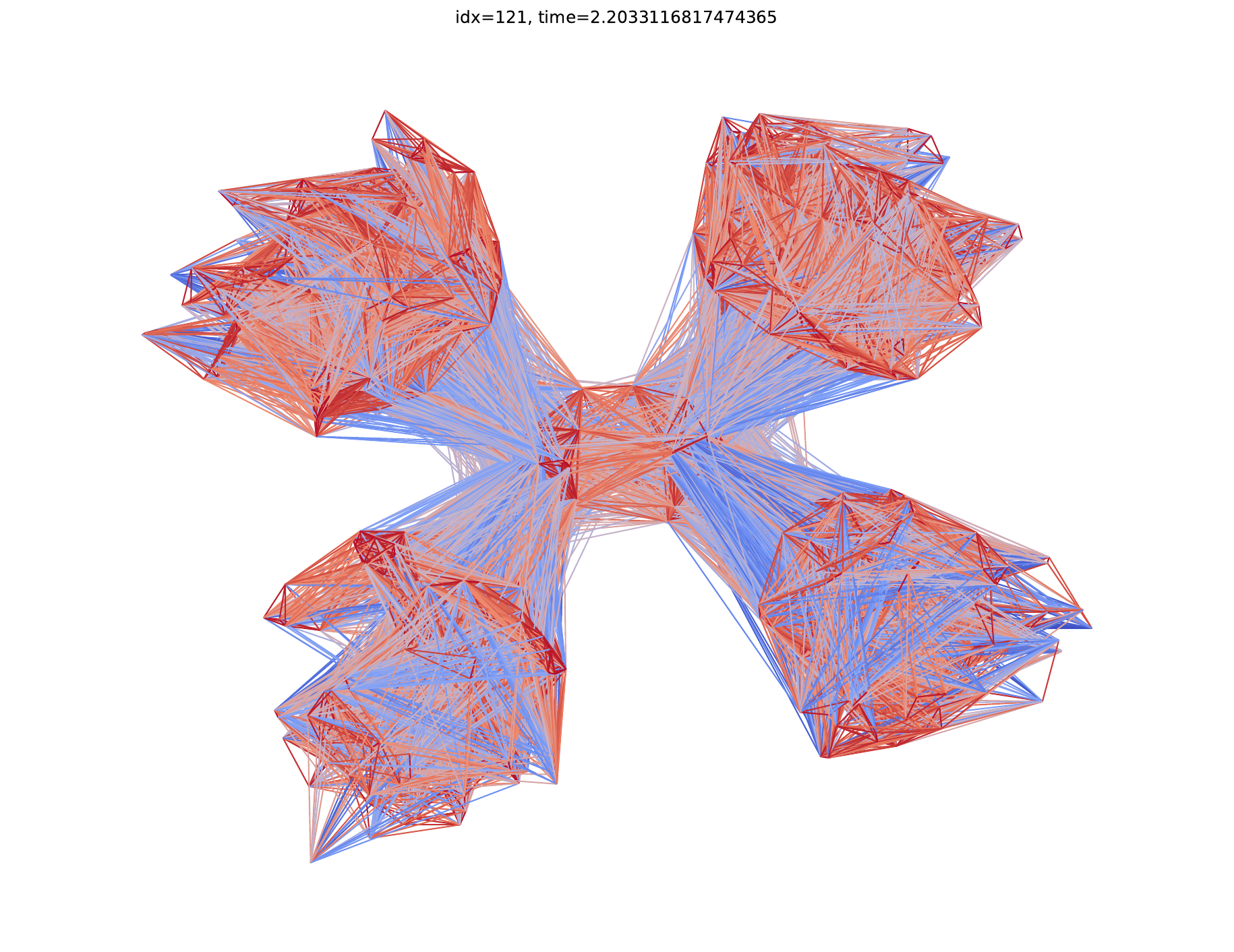} &
\imgcell{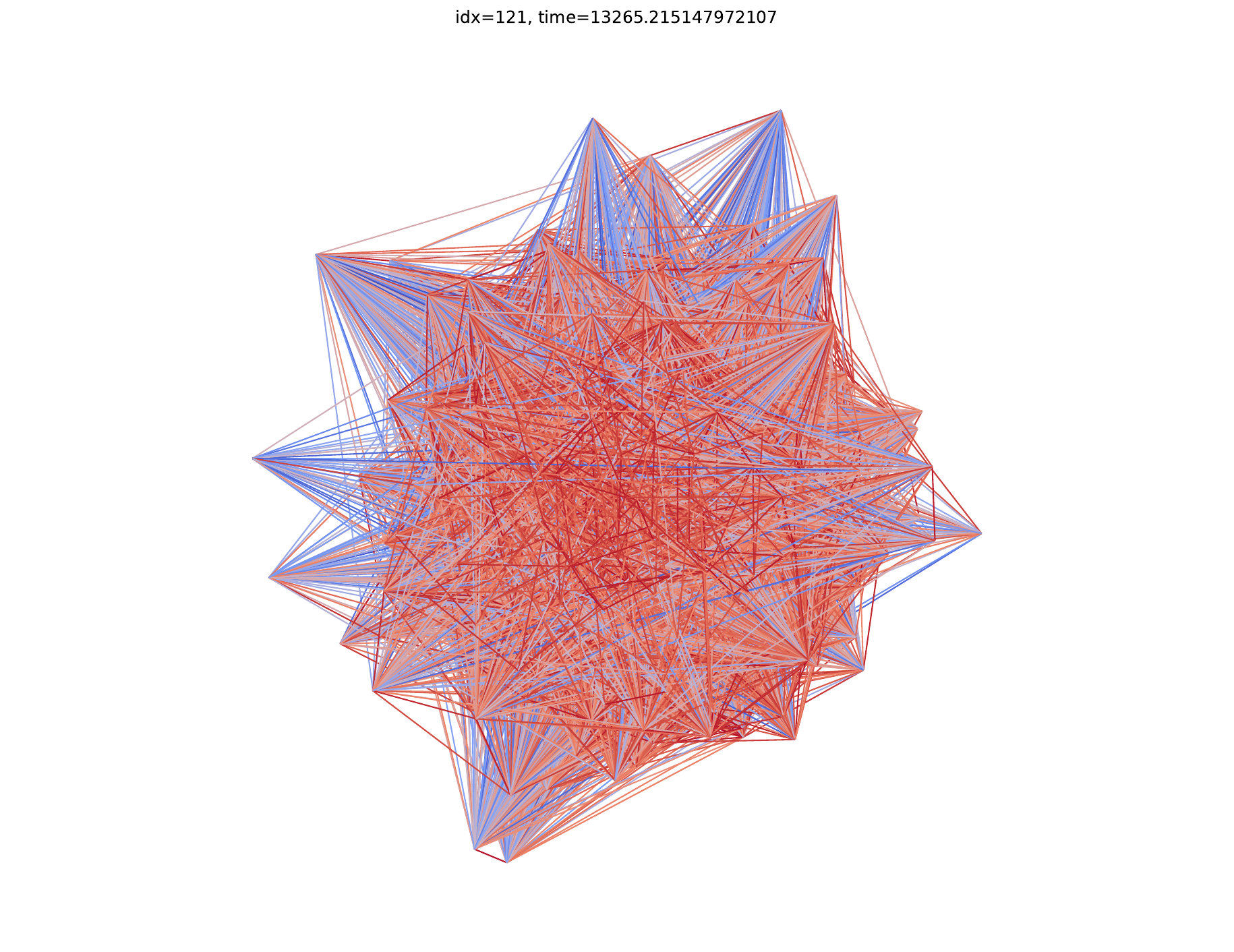} &
\imgcell{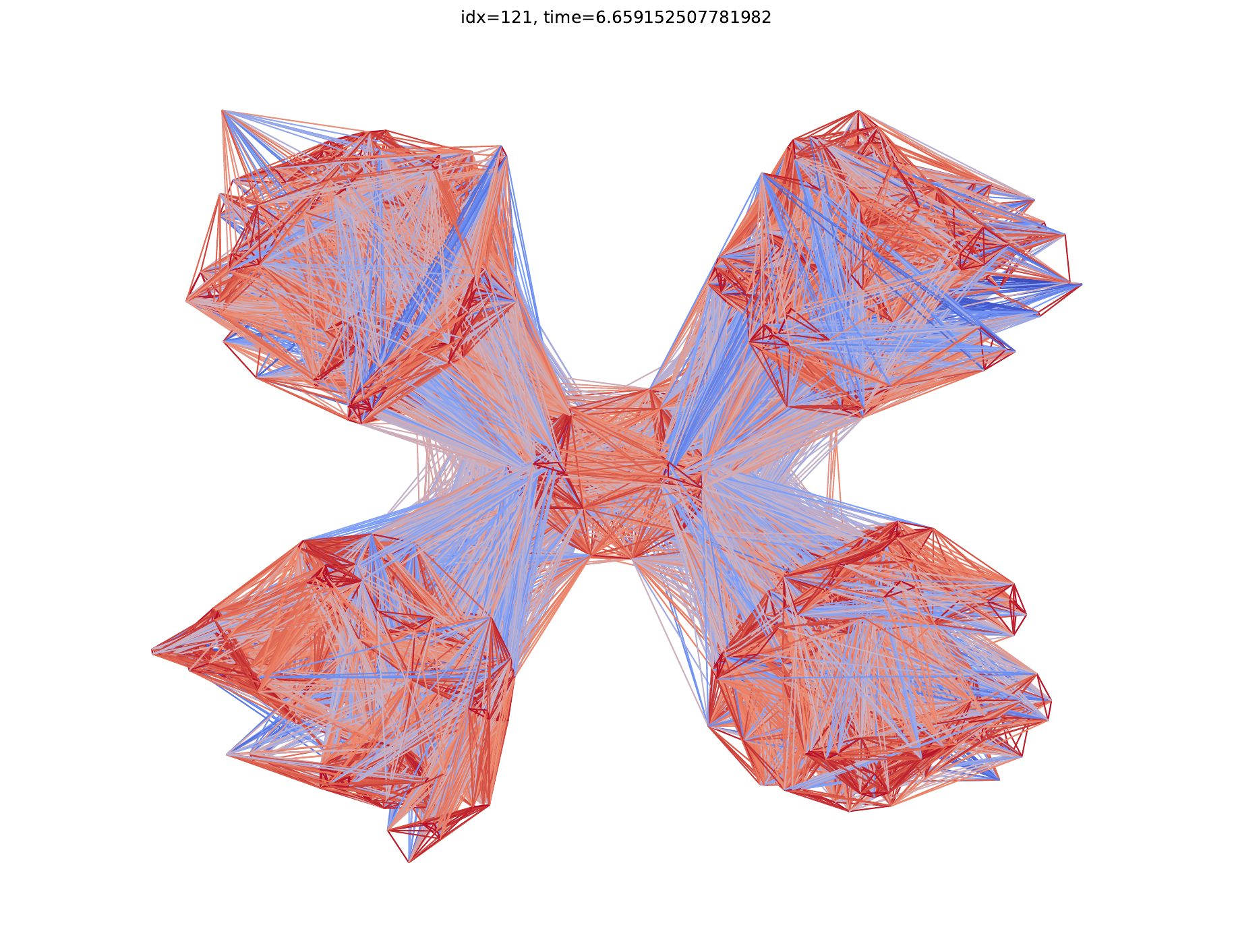} &
\imgcell{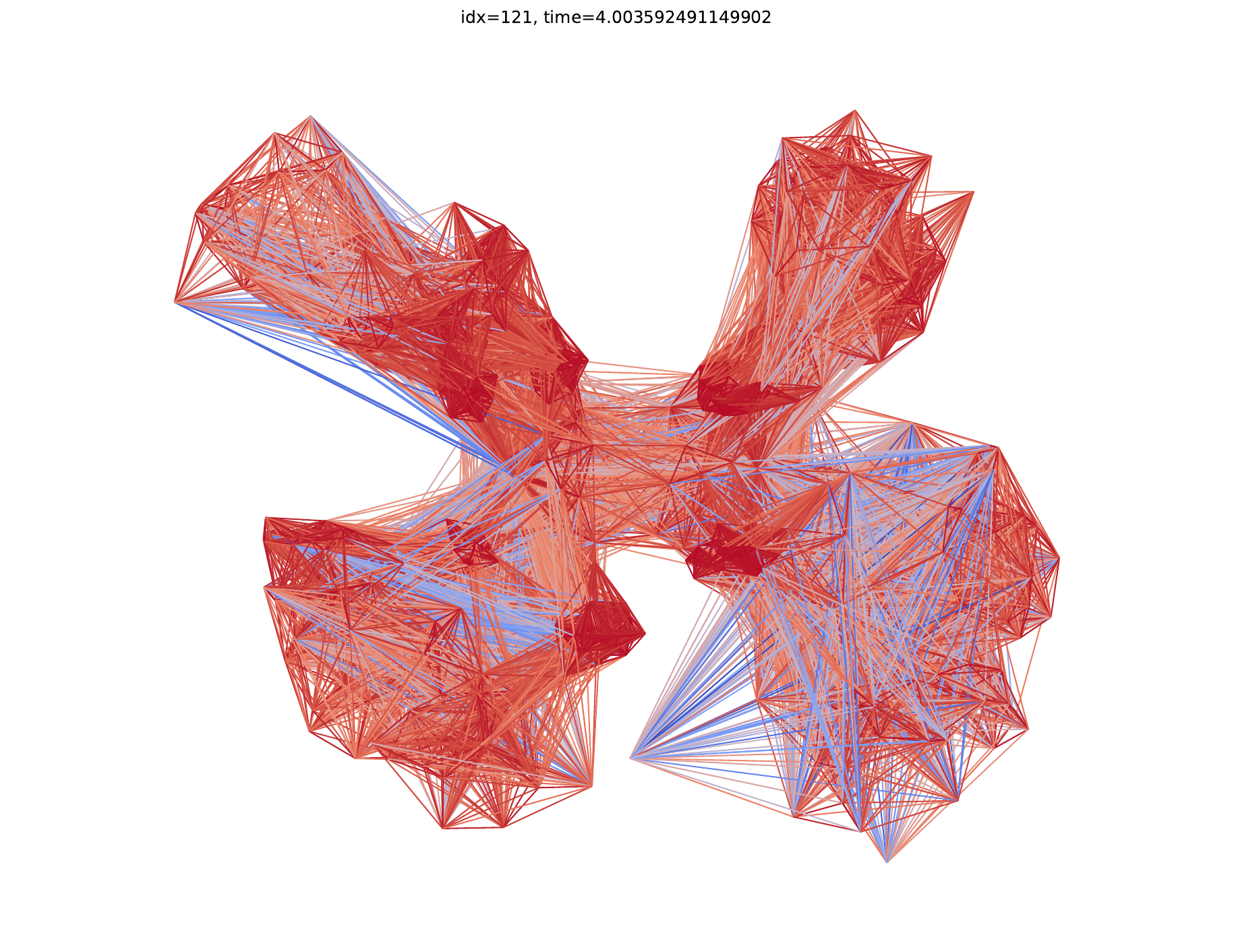} &
\imgcell{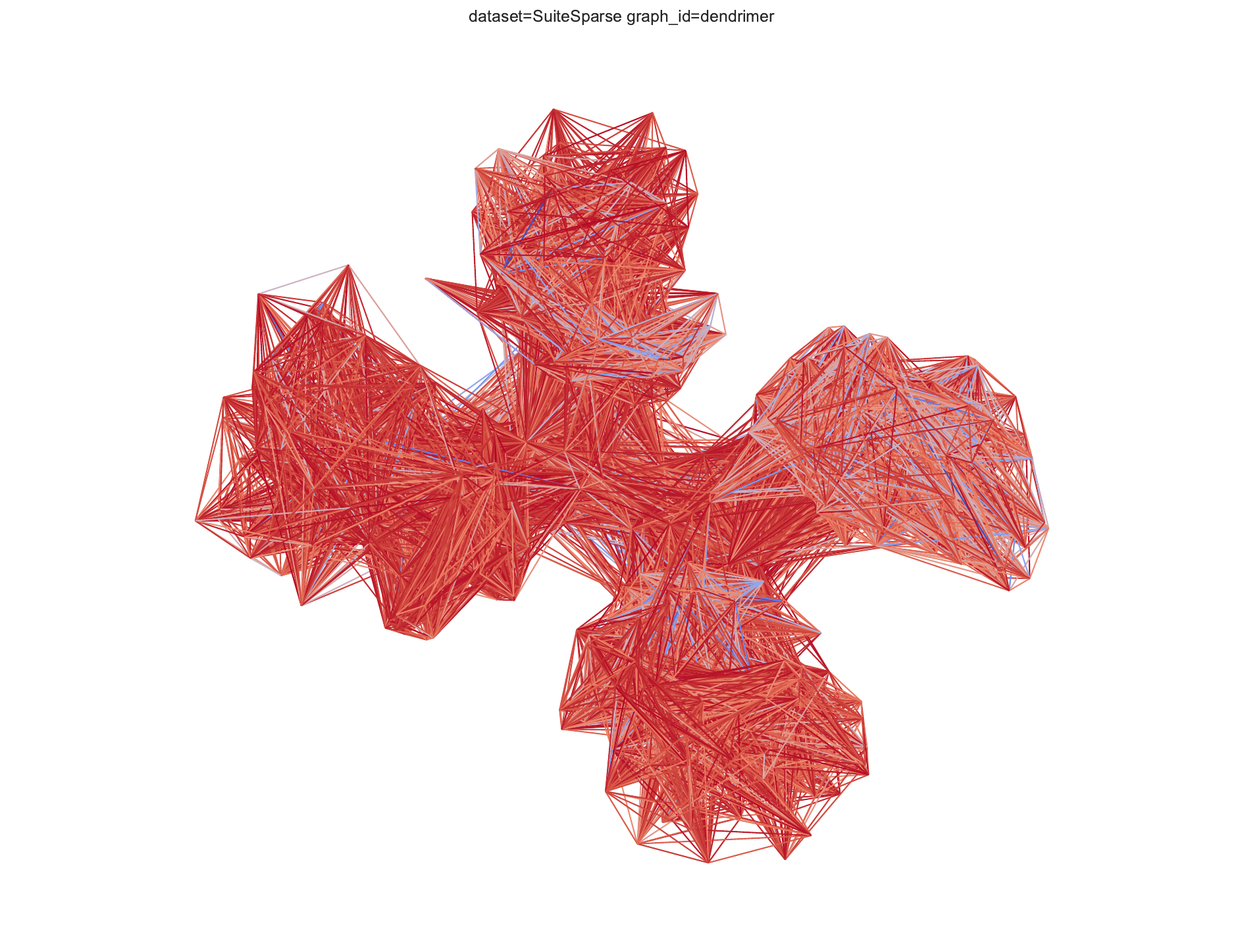} &
\imgcell{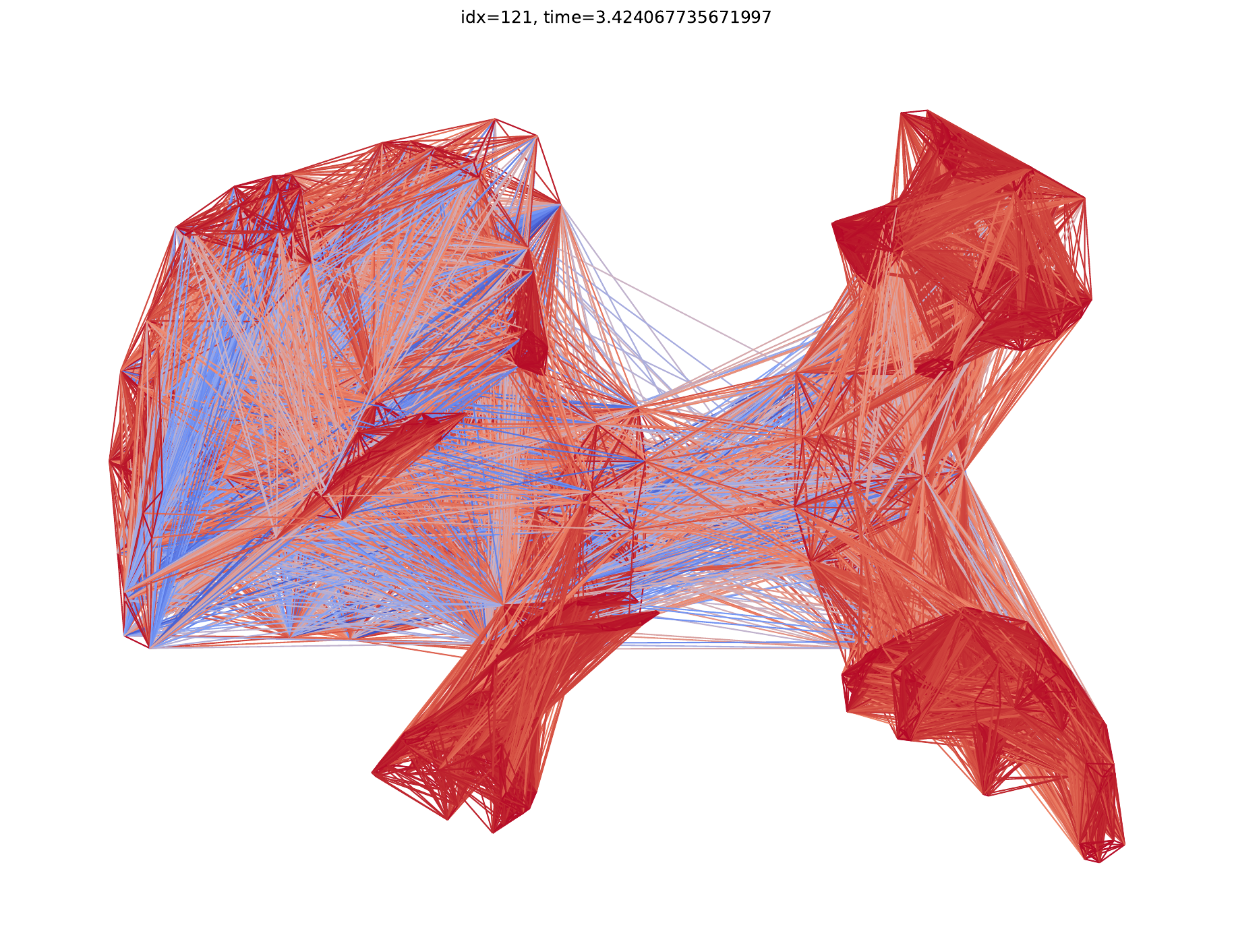} &
\imgcell{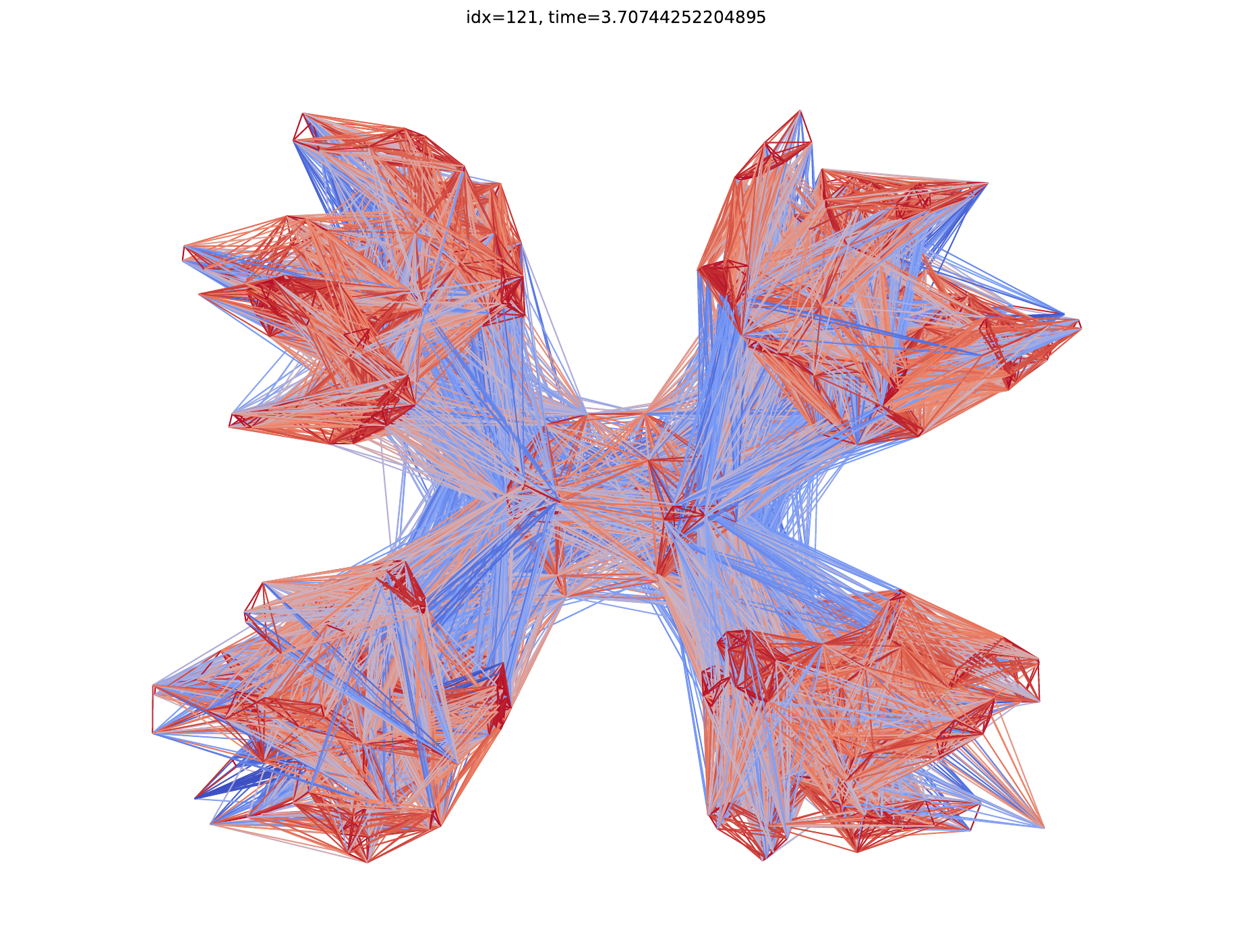} &
\imgcell{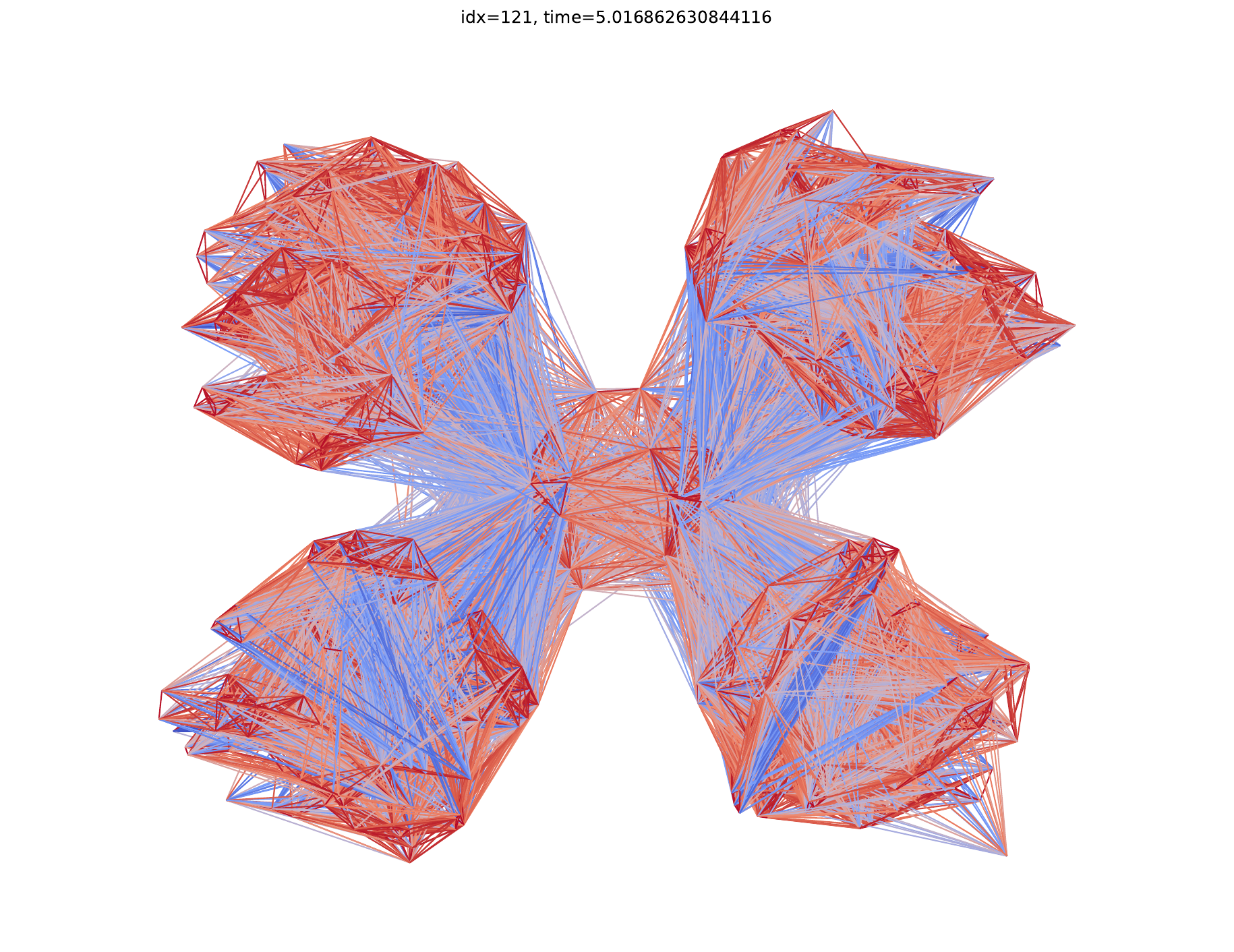} &
\imgcell{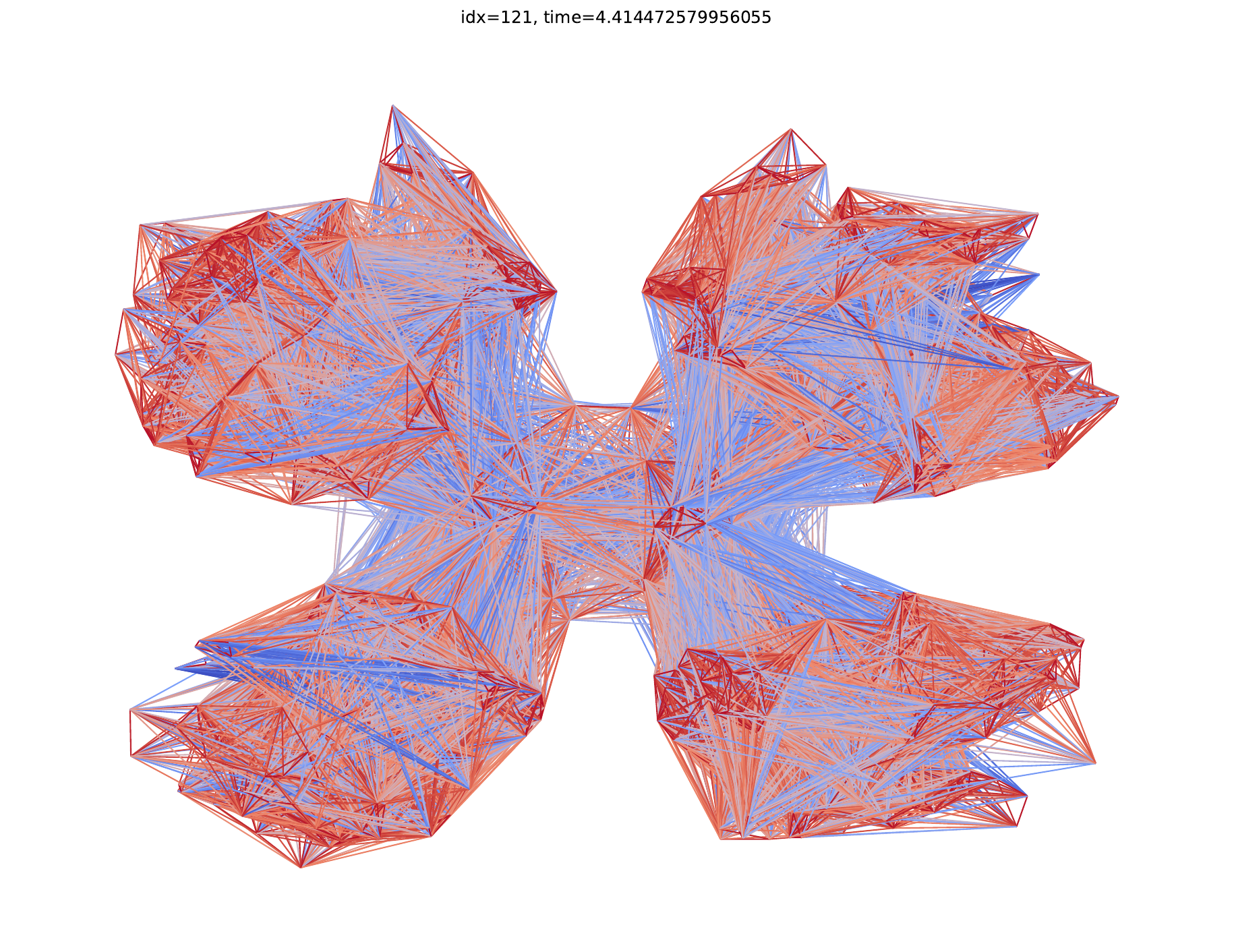} \\

&
t = 0.27s &
t = 103.06s &
t = 44.01s &
t = 2.20s &
t = 7200.00s &
t = 2.24s &
t = 2.05s &
t = 1.99s &
t = 2.42s &
t = 1.85s &
t = 2.46s &
t = 1.78s \\

\makecell{\bfseries Si2\\N = 769\\M = 8516} &
\imgcell{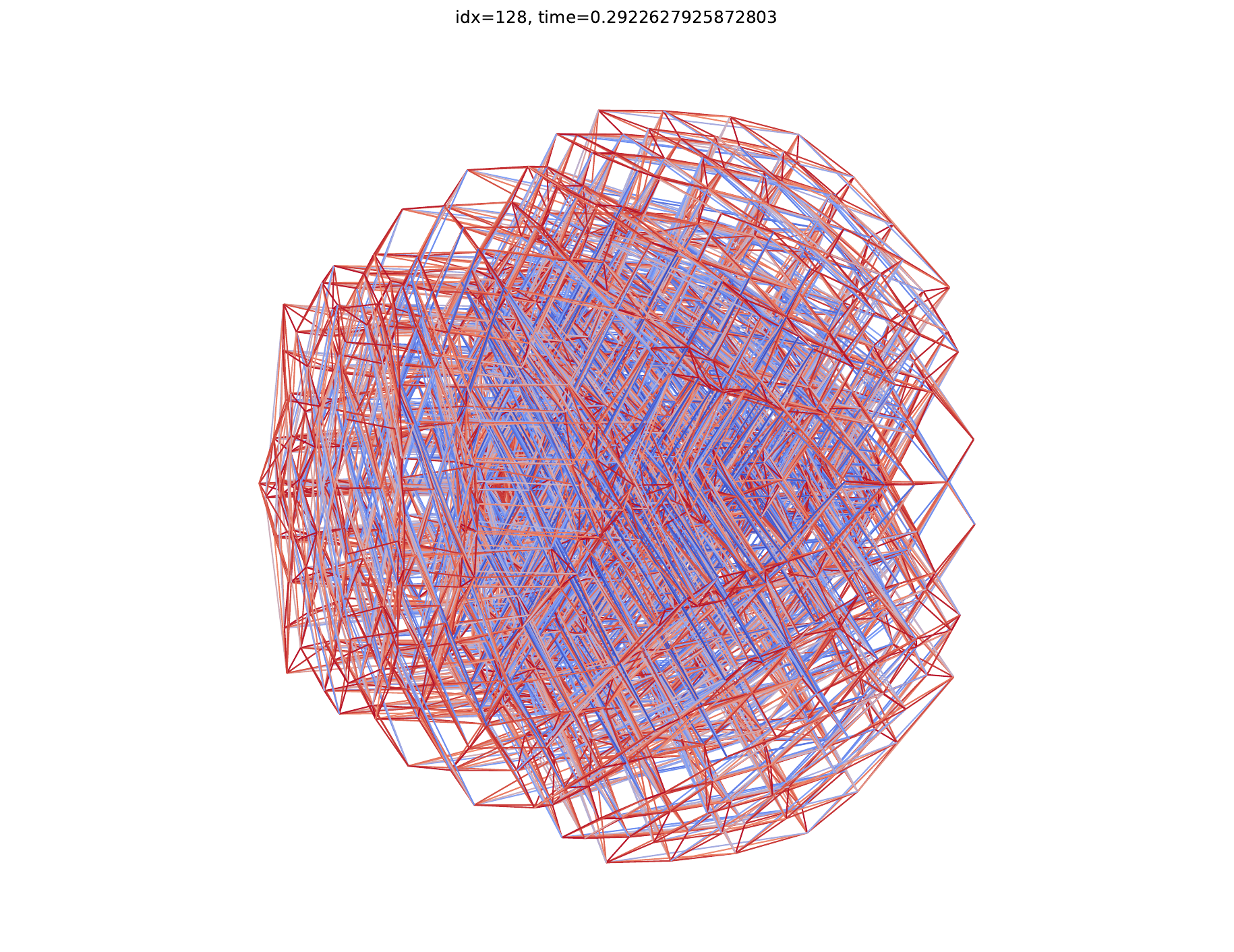} &
\imgcell{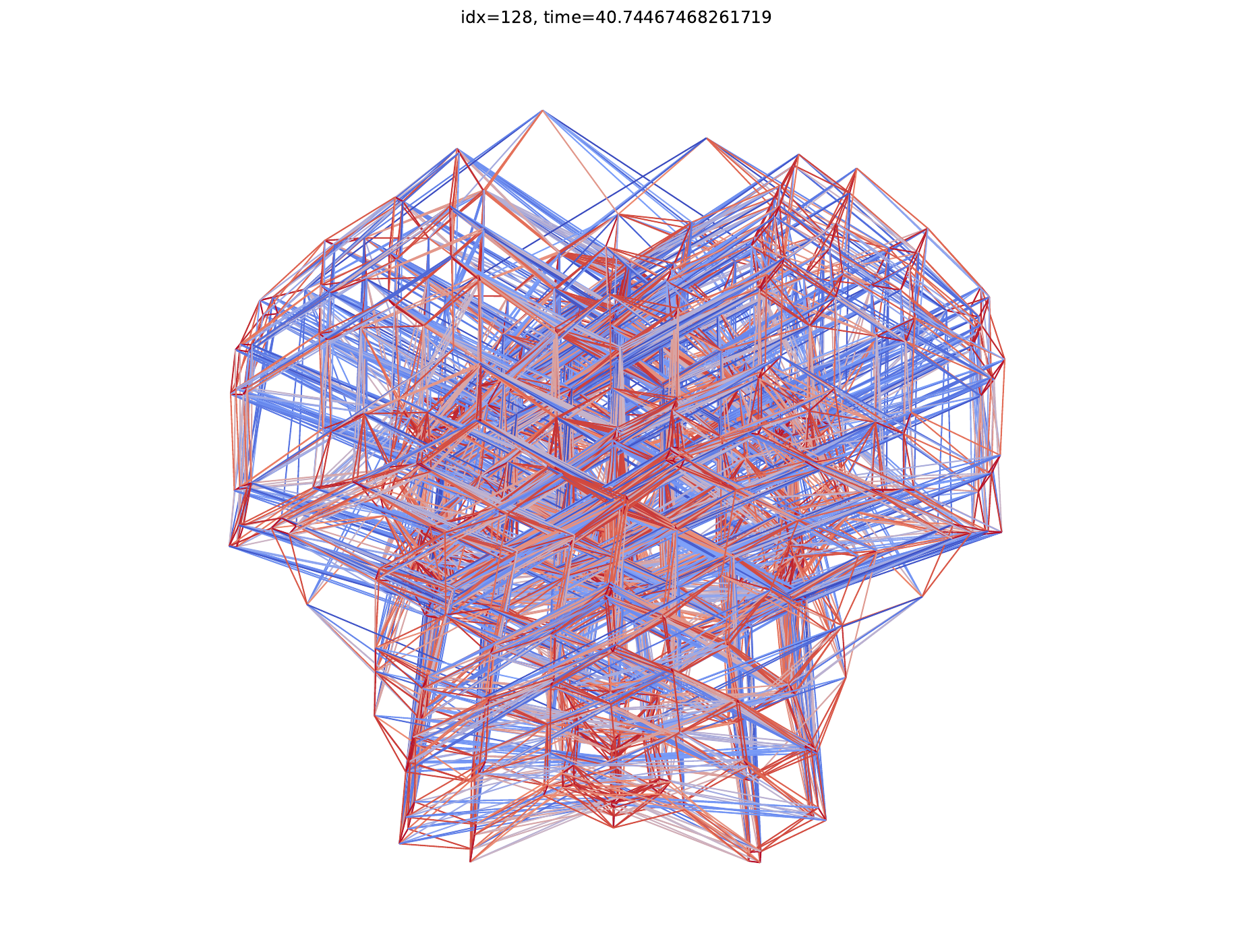} &
\imgcell{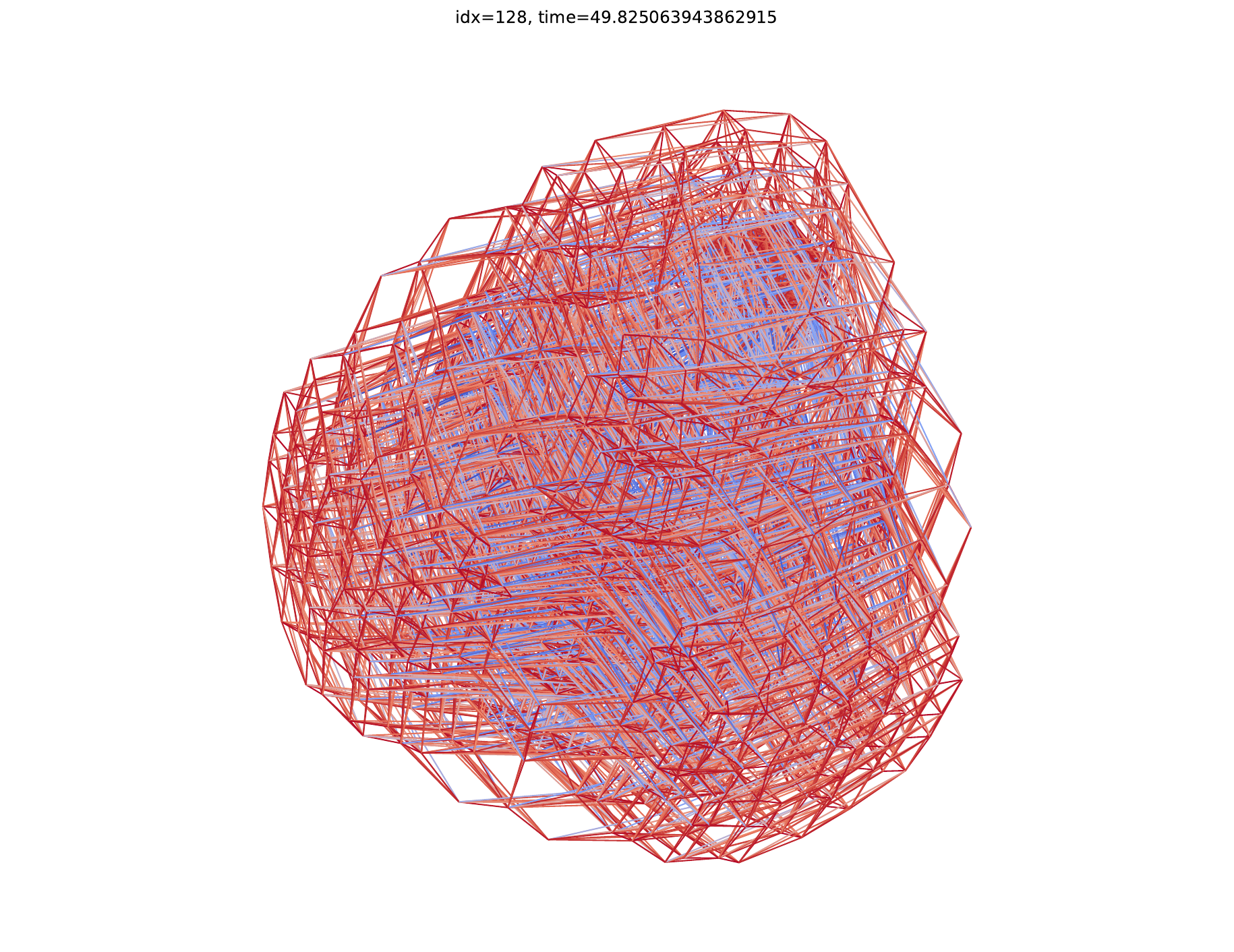} &
\imgcell{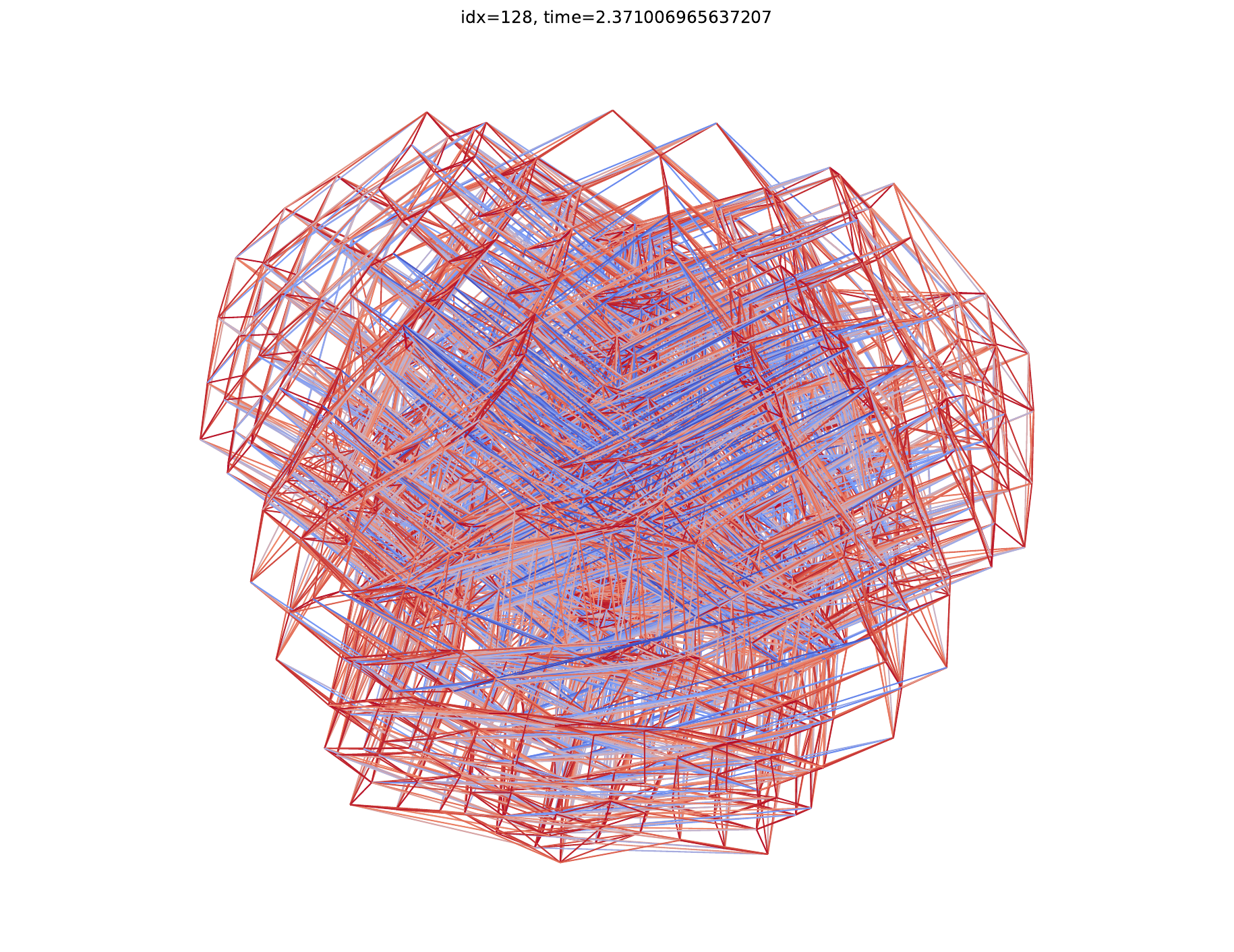} &
\imgcell{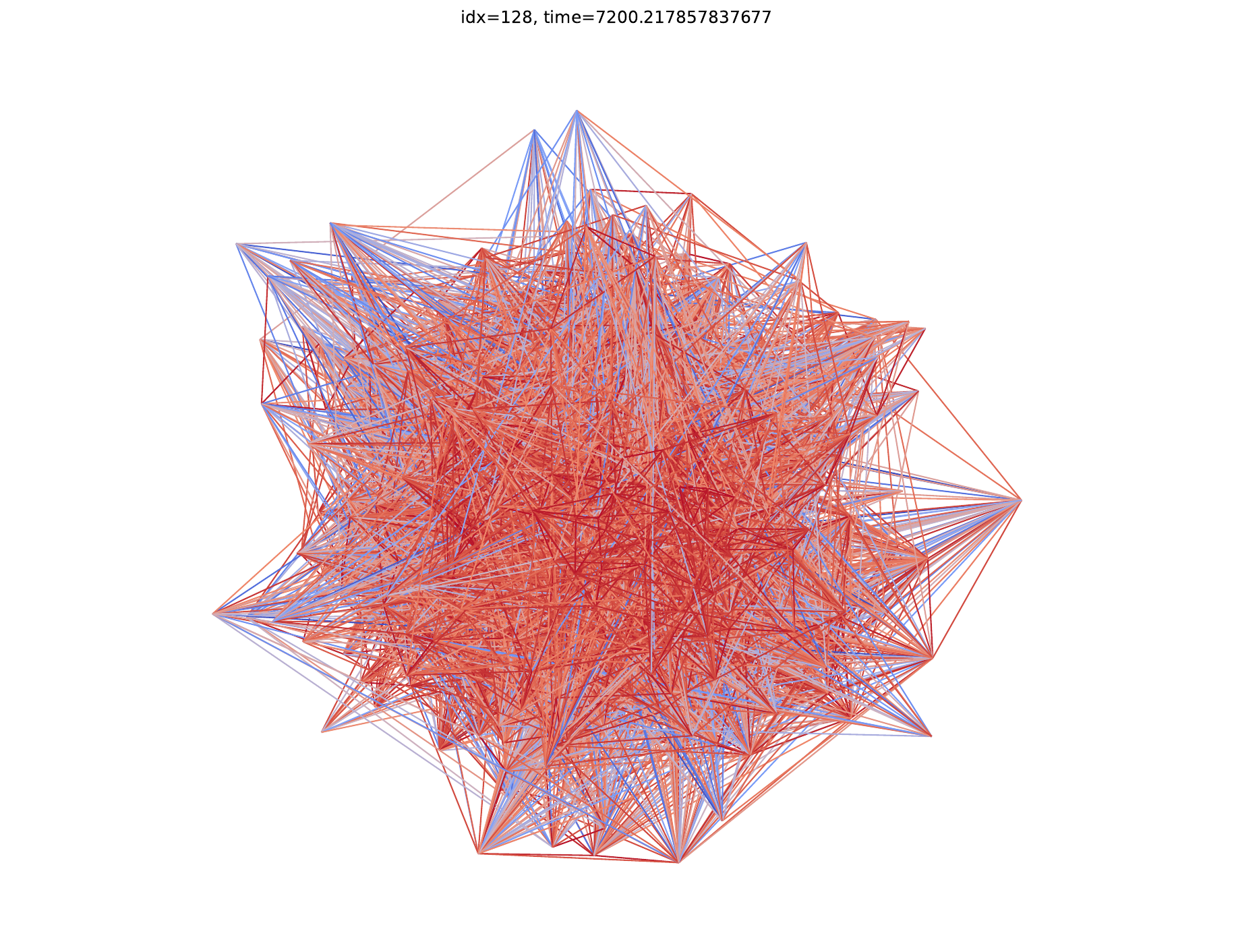} &
\imgcell{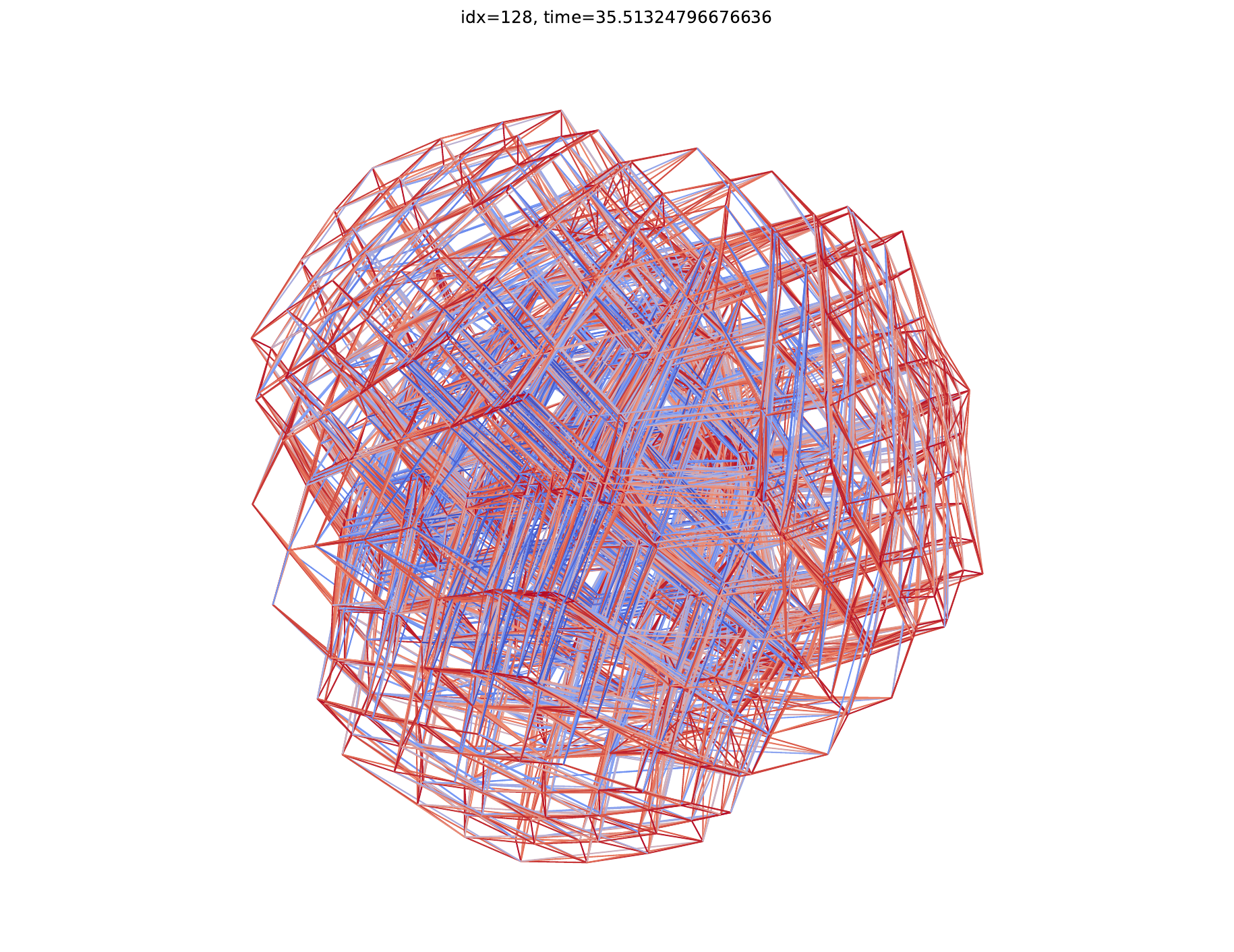} &
\imgcell{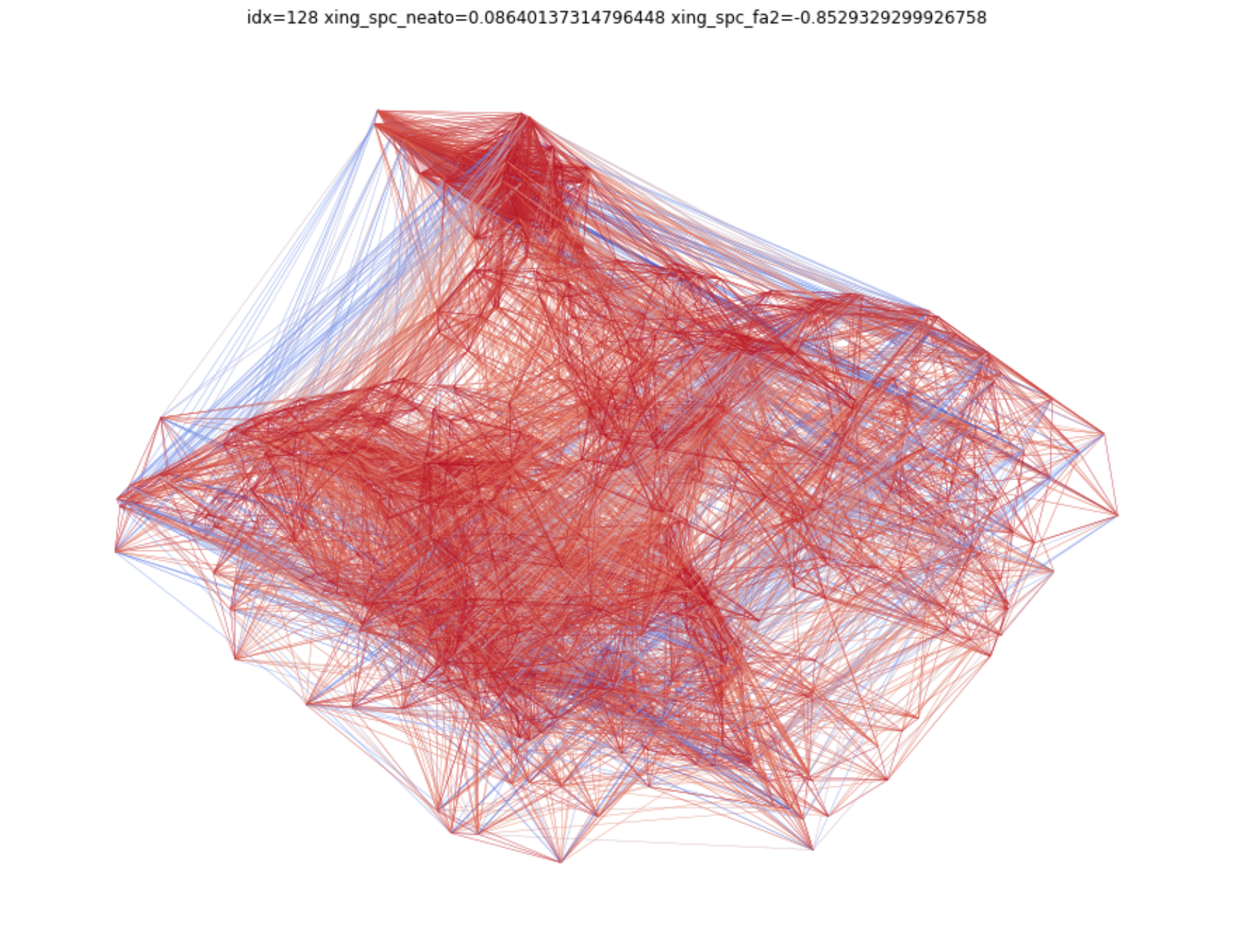} &
\imgcell{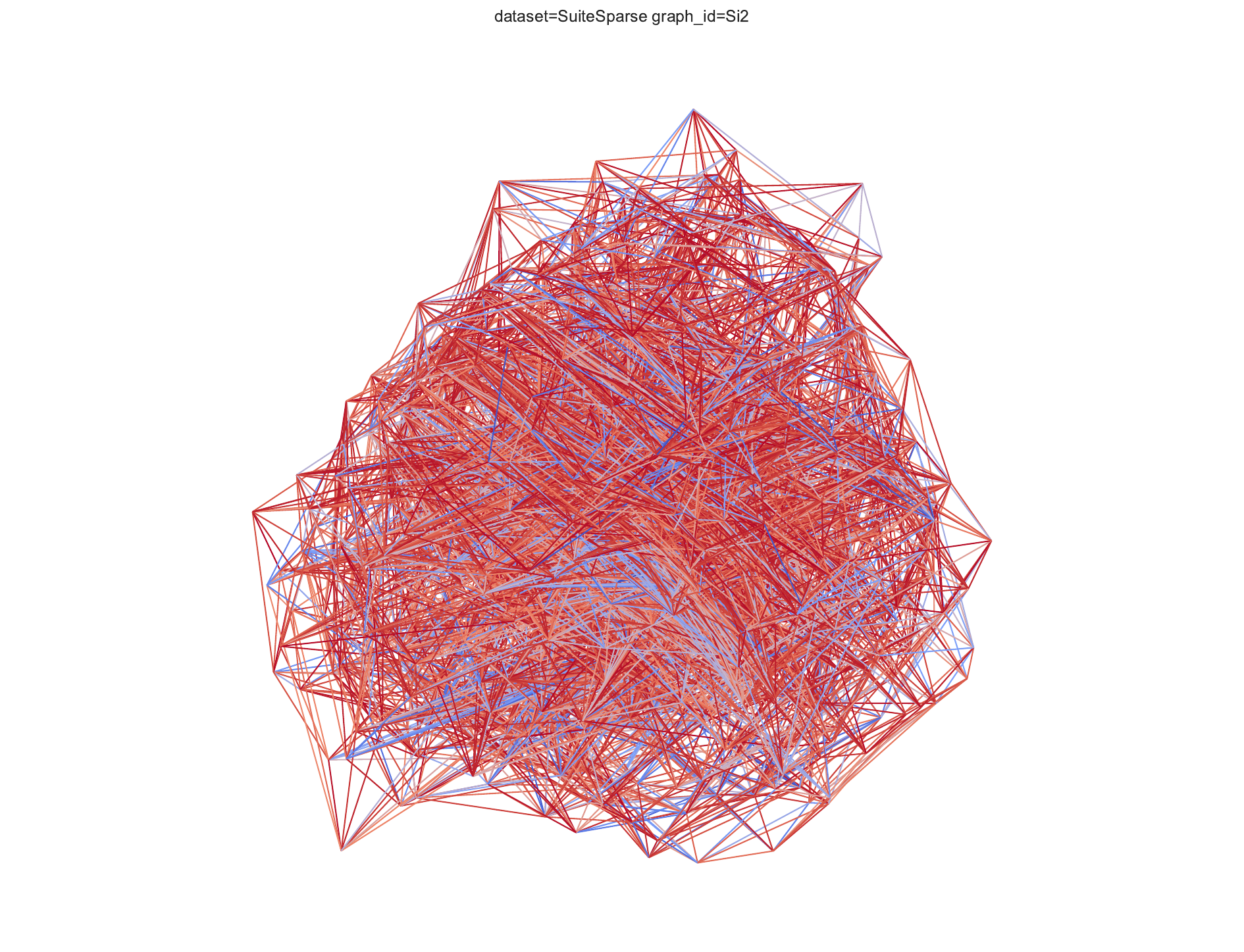} &
\imgcell{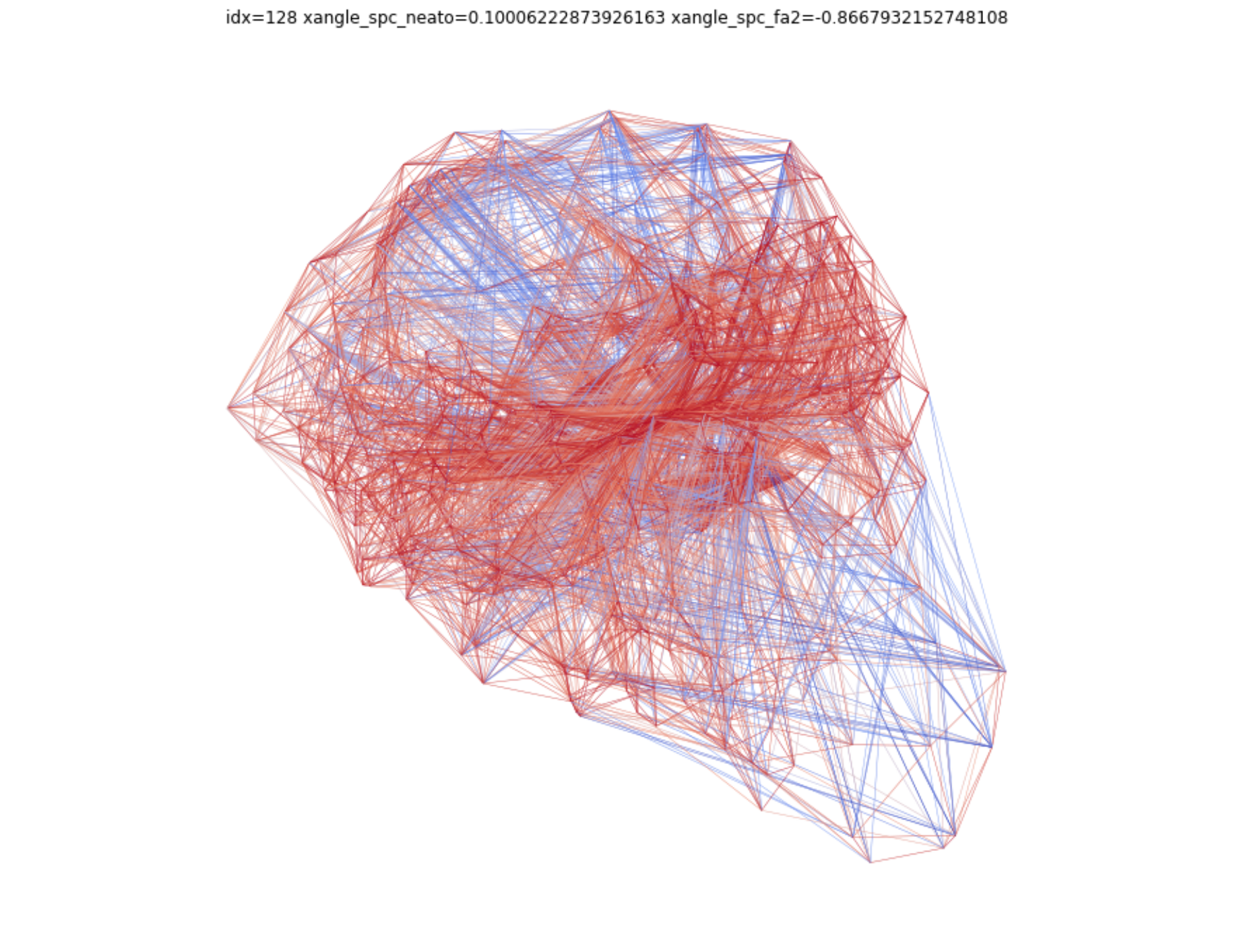} &
\imgcell{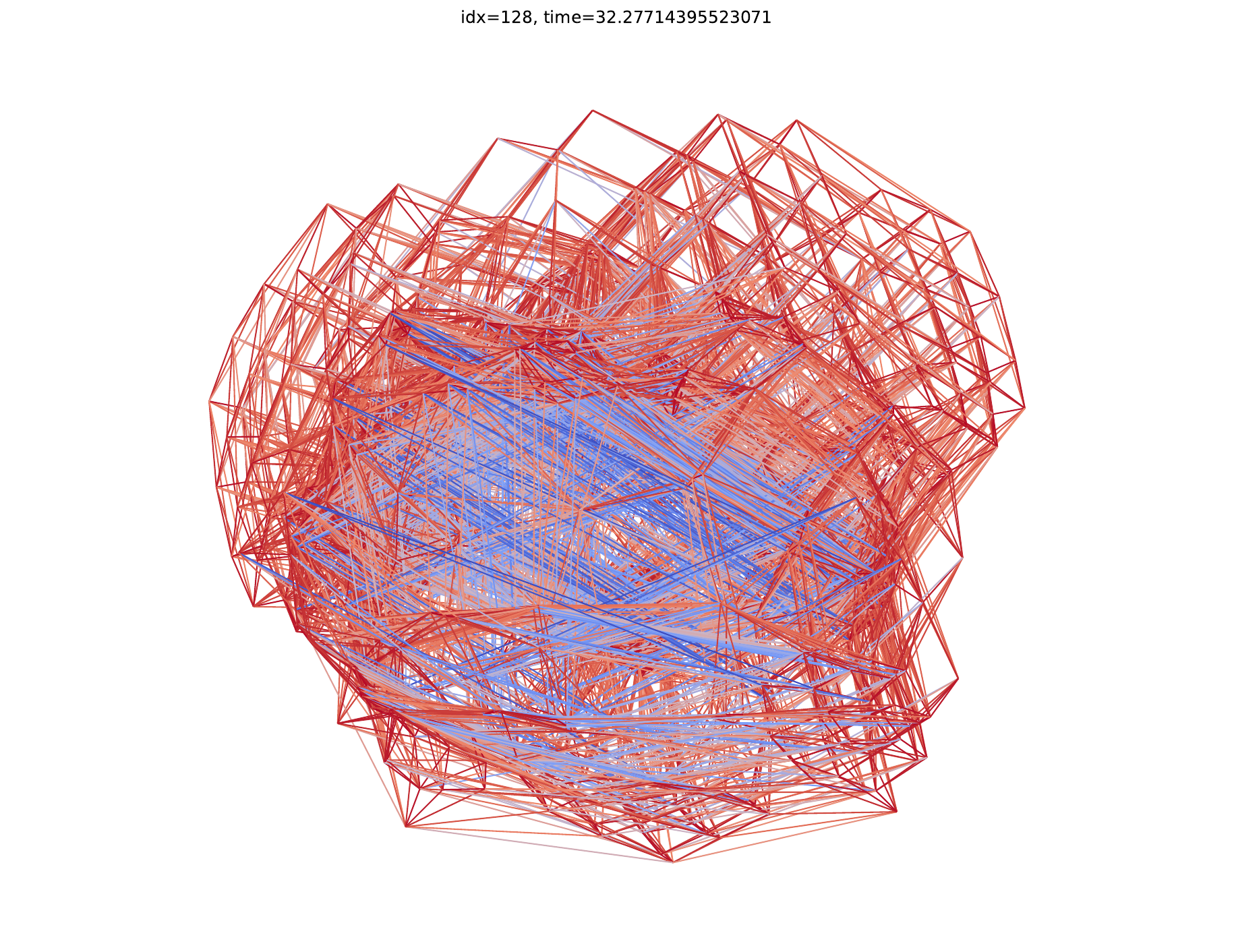} &
\imgcell{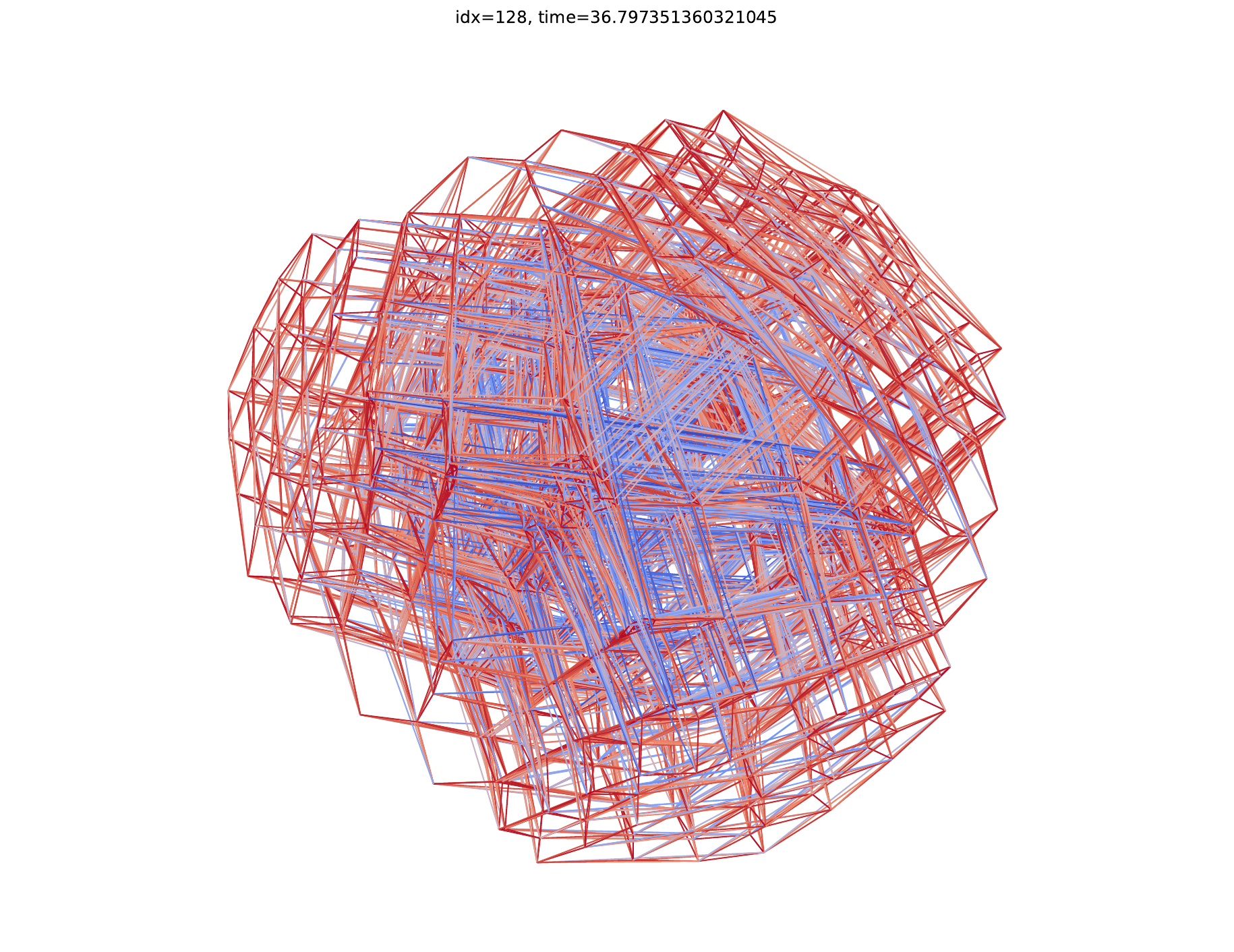} &
\imgcell{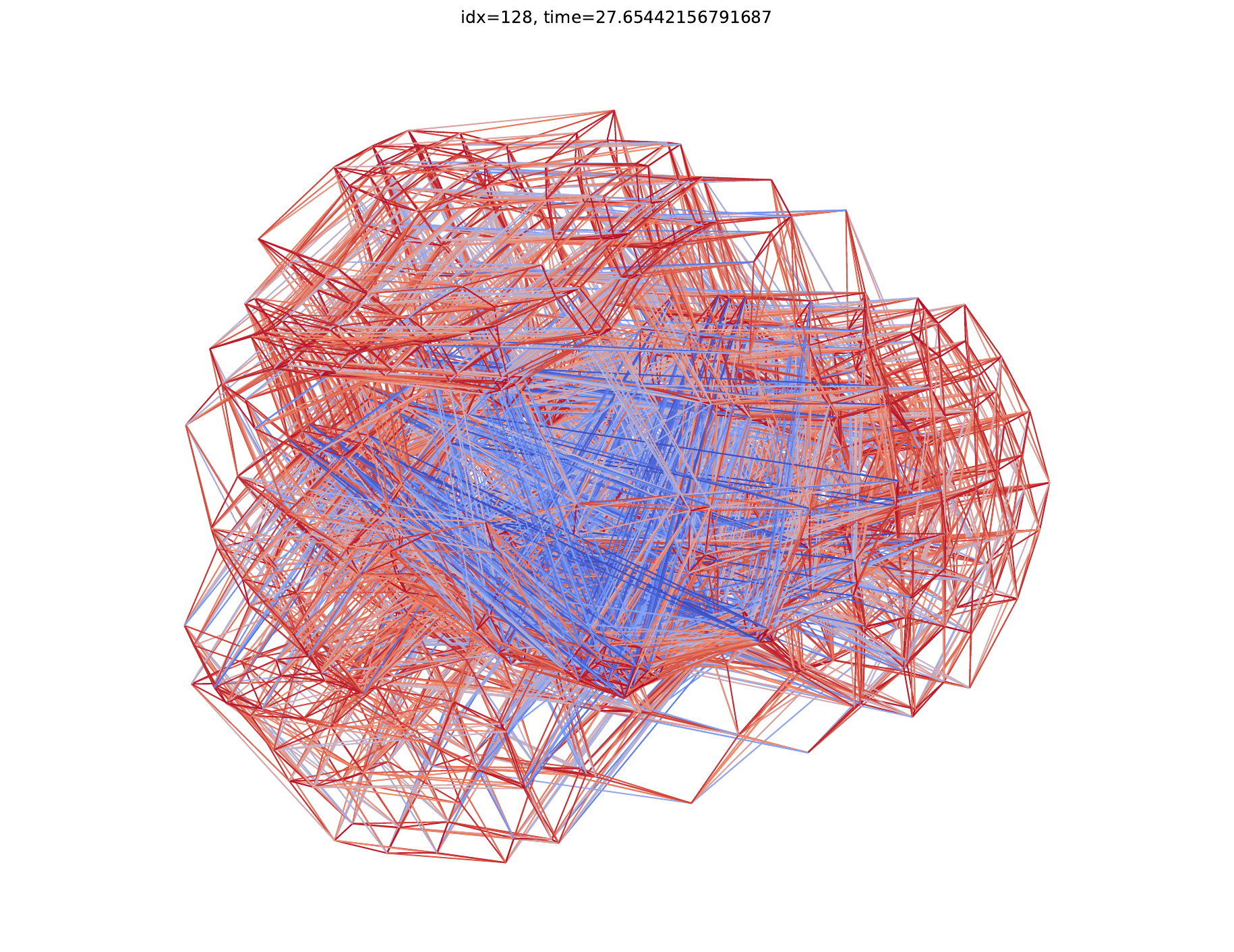} \\

&
t = 0.29s &
t = 40.74s &
t = 49.83s &
t = 2.37s &
t = 7200.00s &
t = 2.51s &
t = 1.96s &
t = 2.10s &
t = 1.87s &
t = 2.28s &
t = 2.30s &
t = 2.45s \\

\makecell{\bfseries lshp\_778\\N = 778\\M = 2247} &
\imgcell{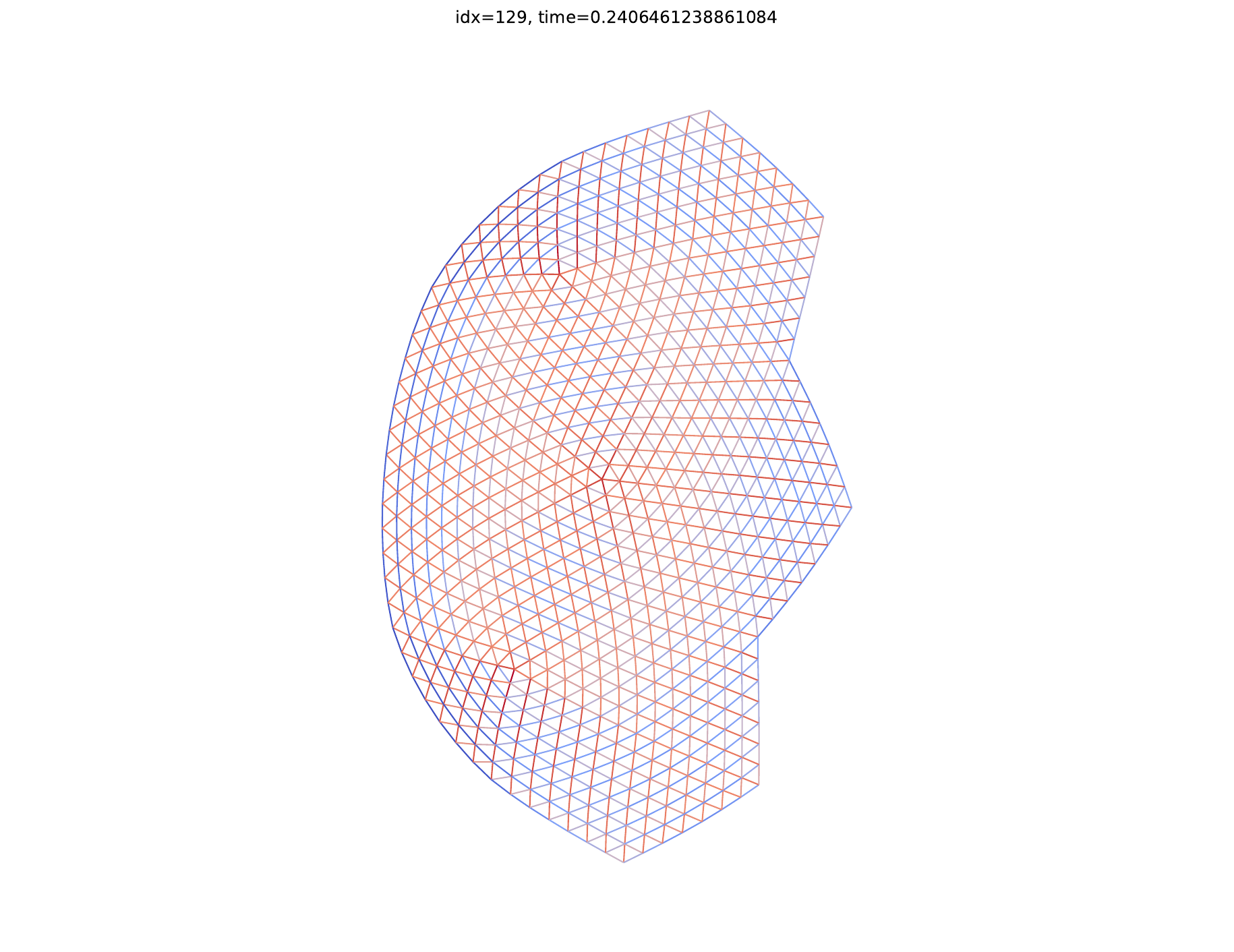} &
\imgcell{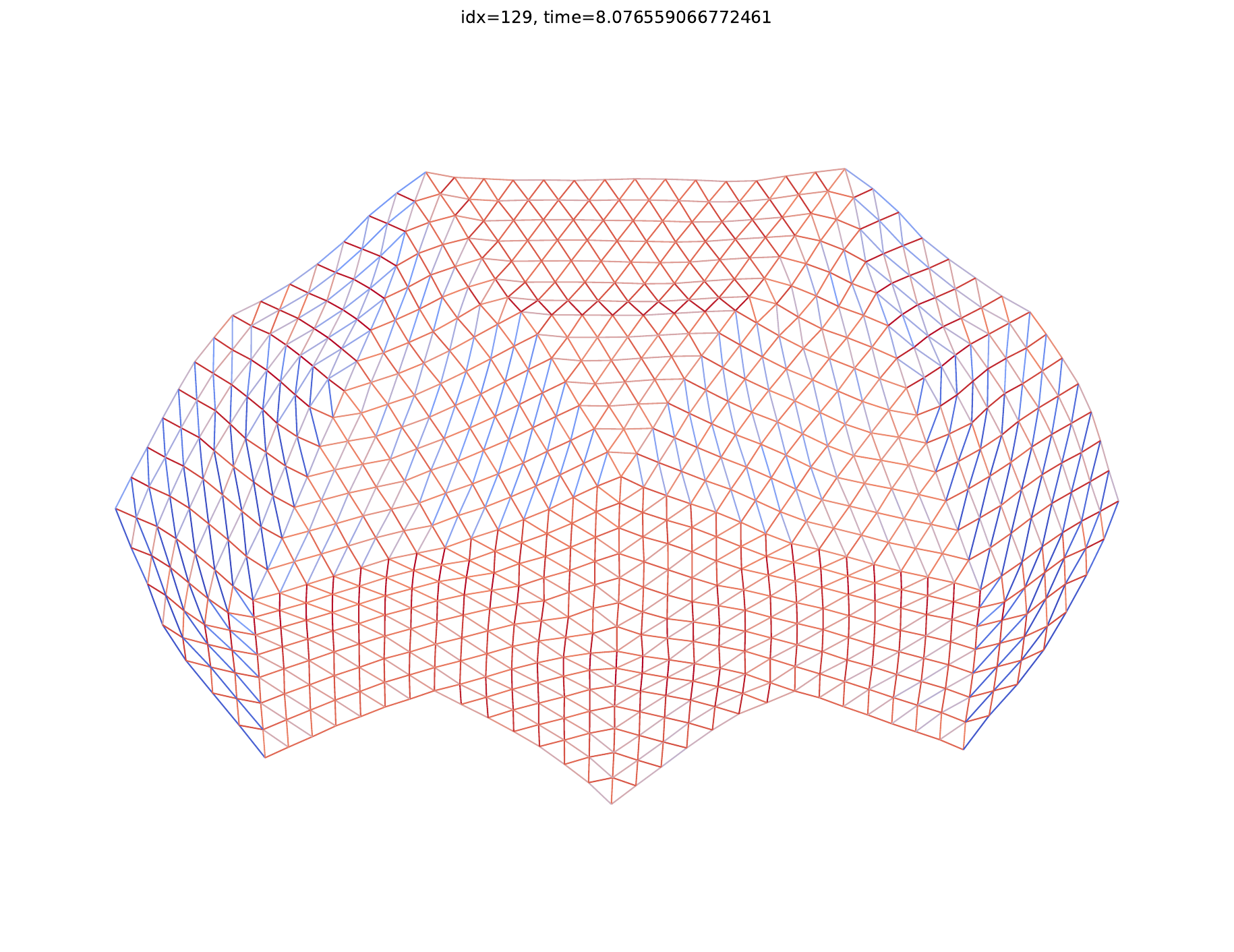} &
\imgcell{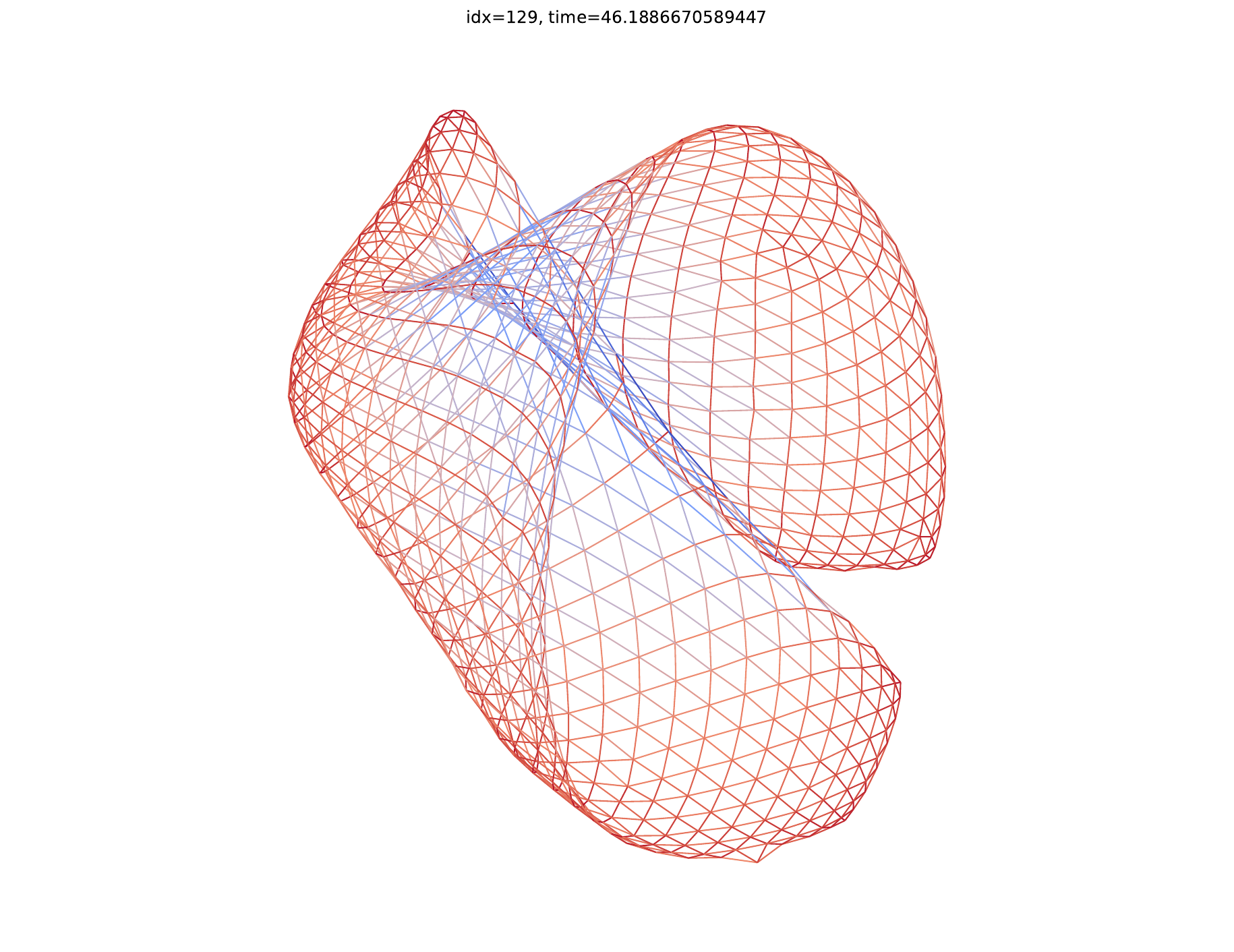} &
\imgcell{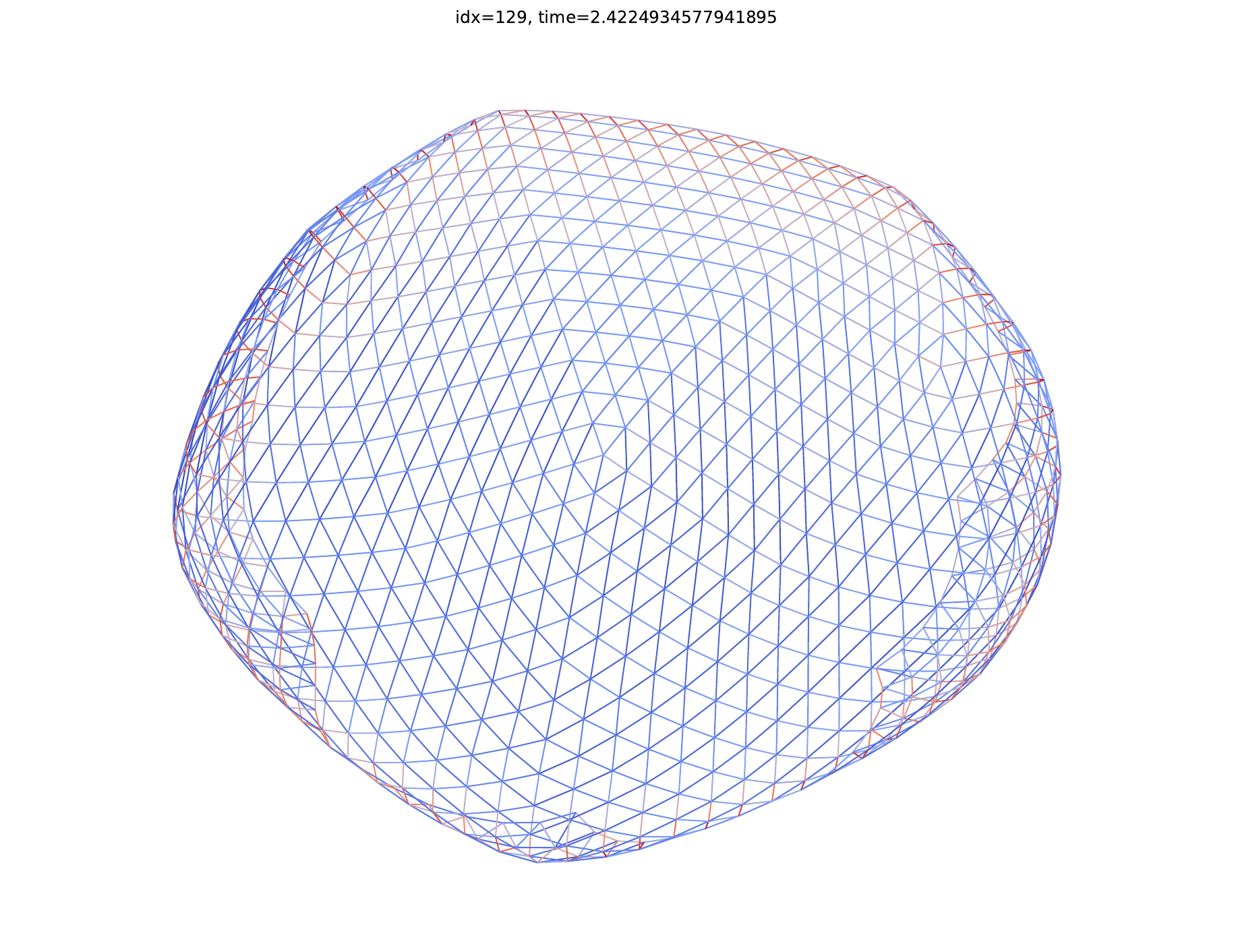} &
\imgcell{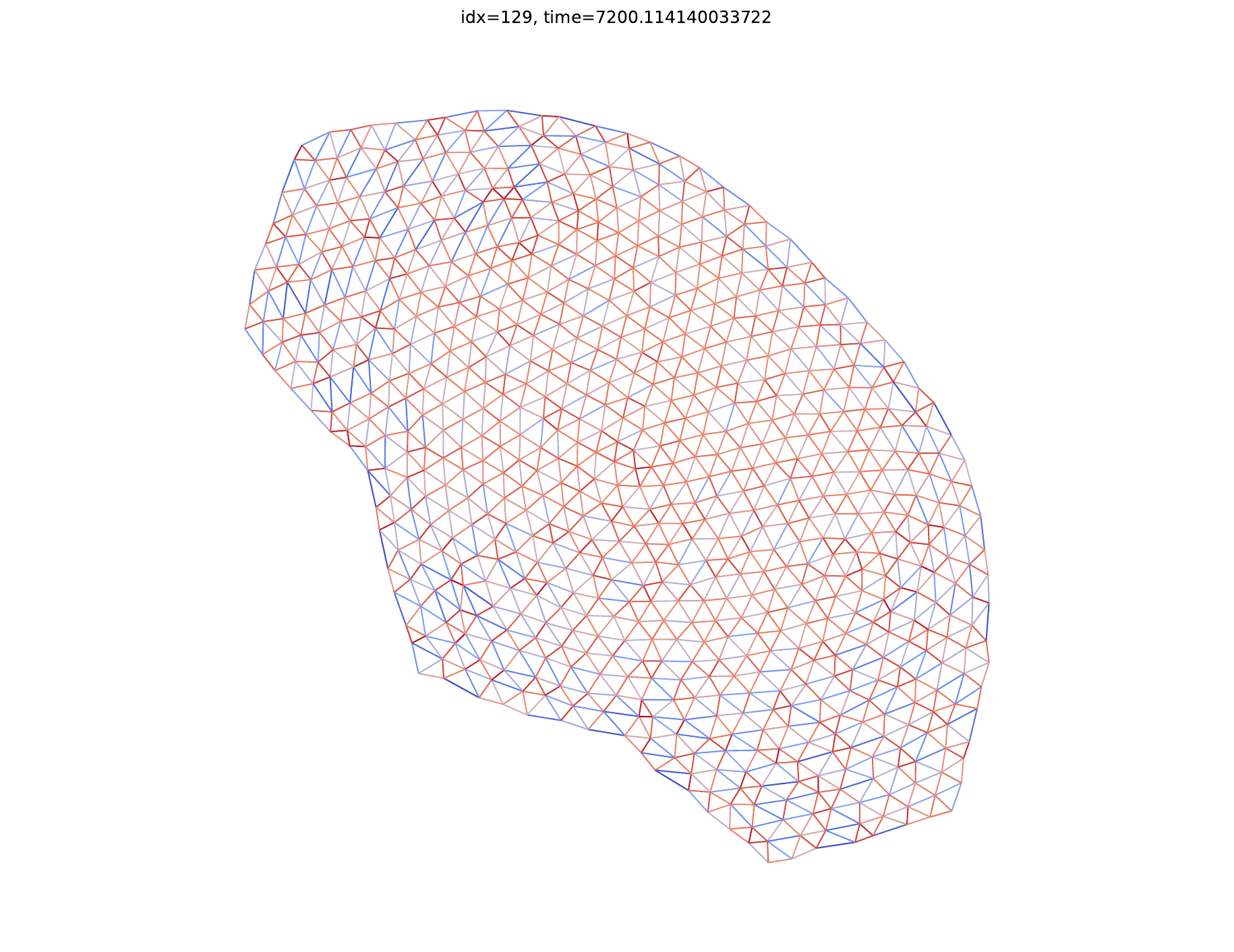} &
\imgcell{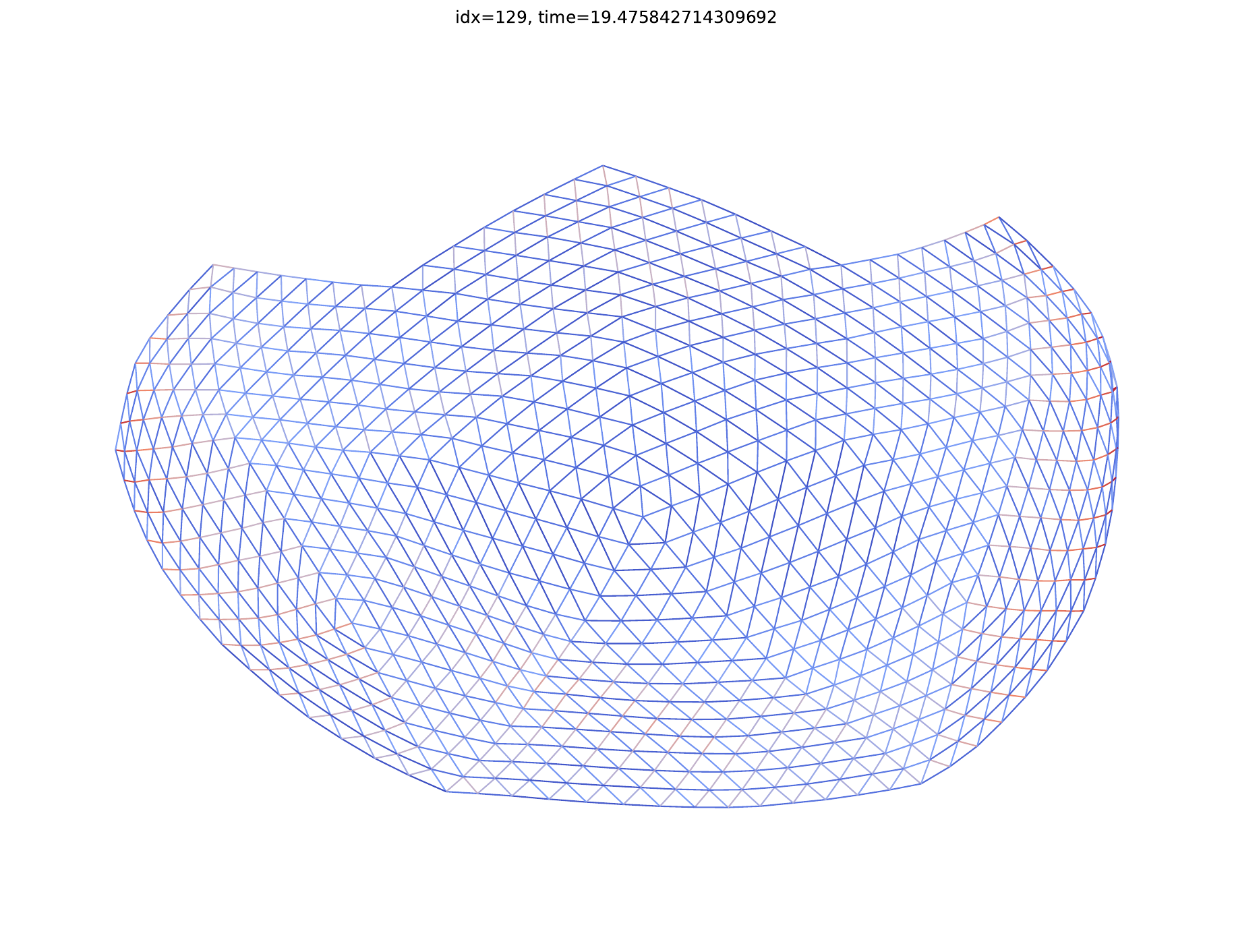} &
\imgcell{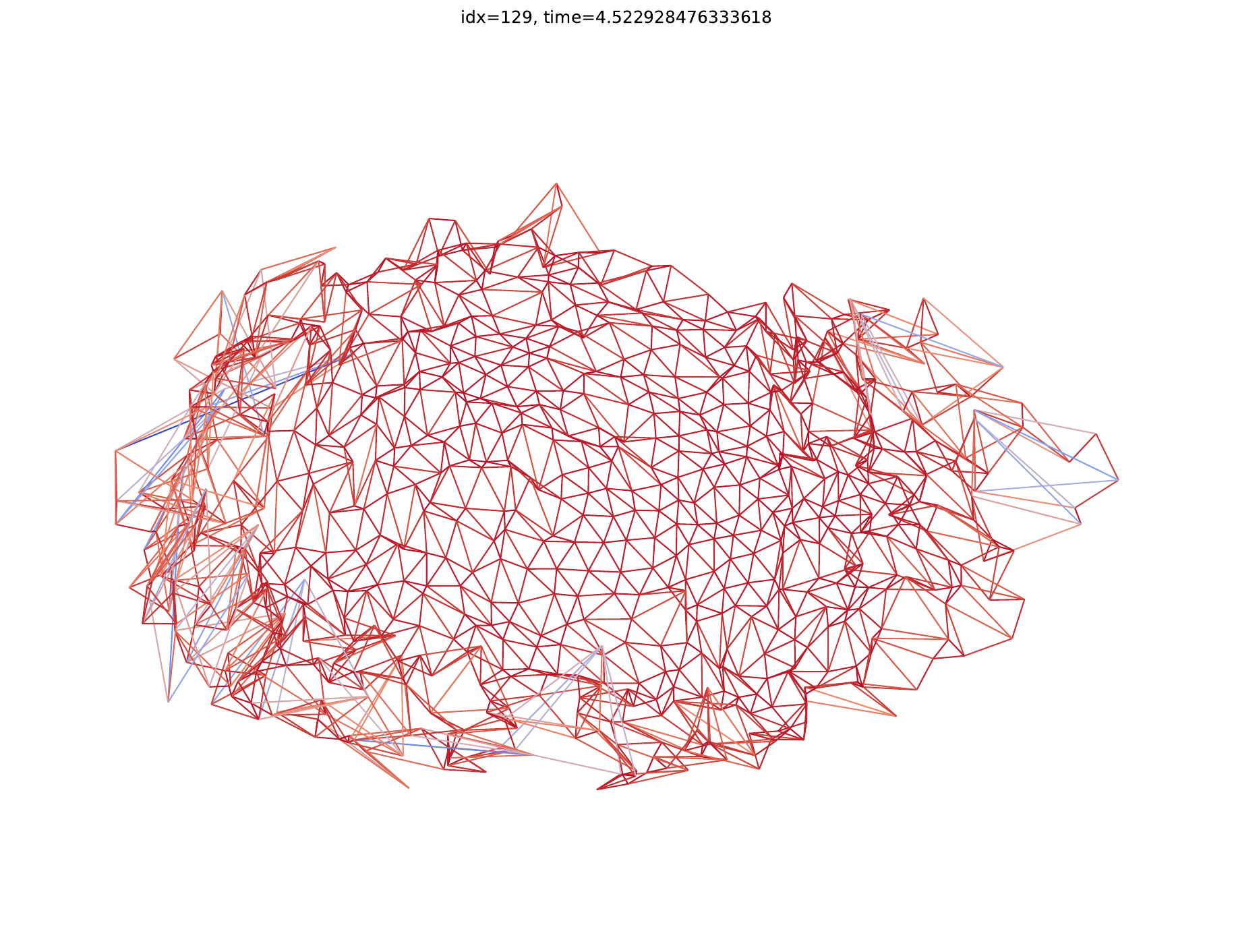} &
\imgcell{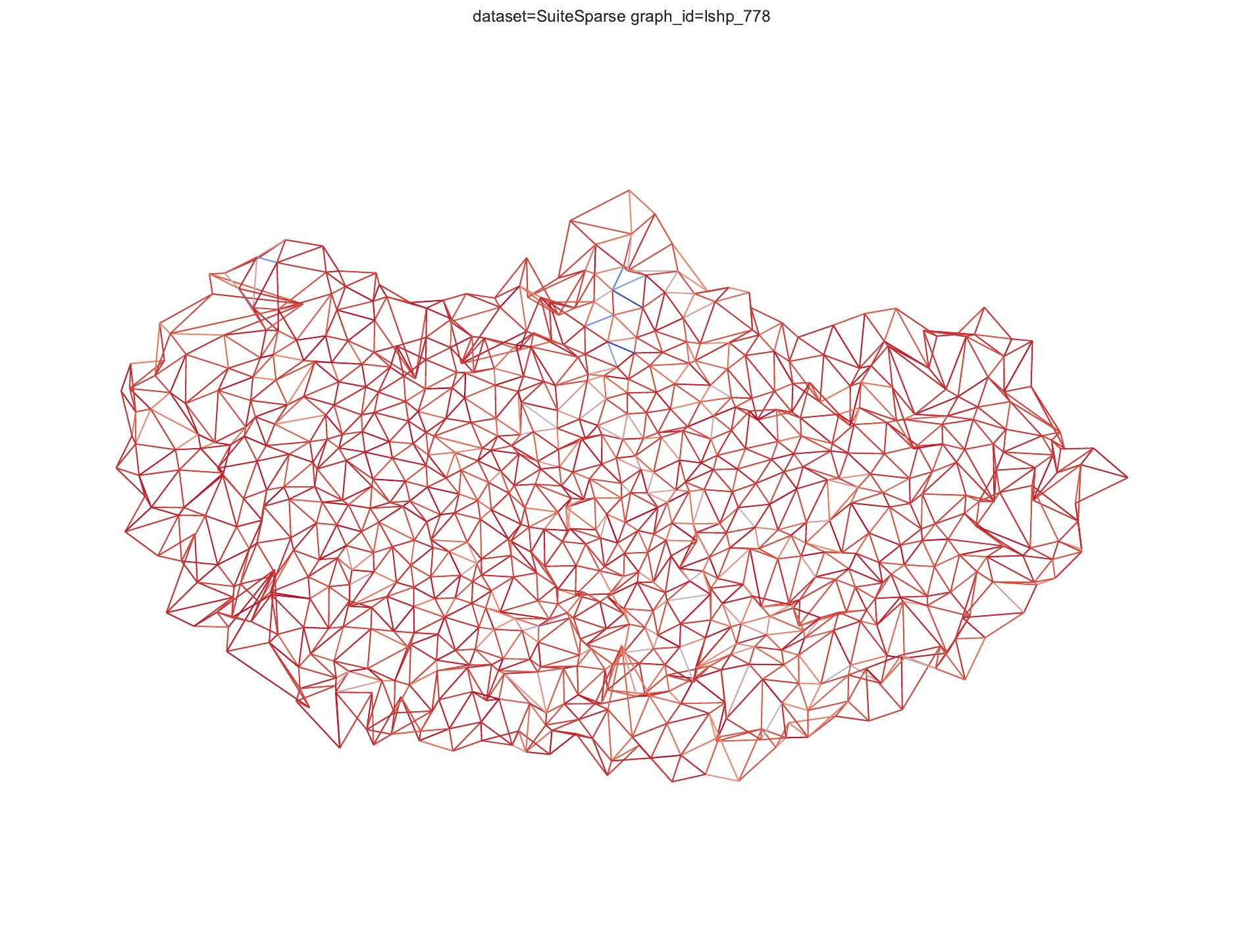} &
\imgcell{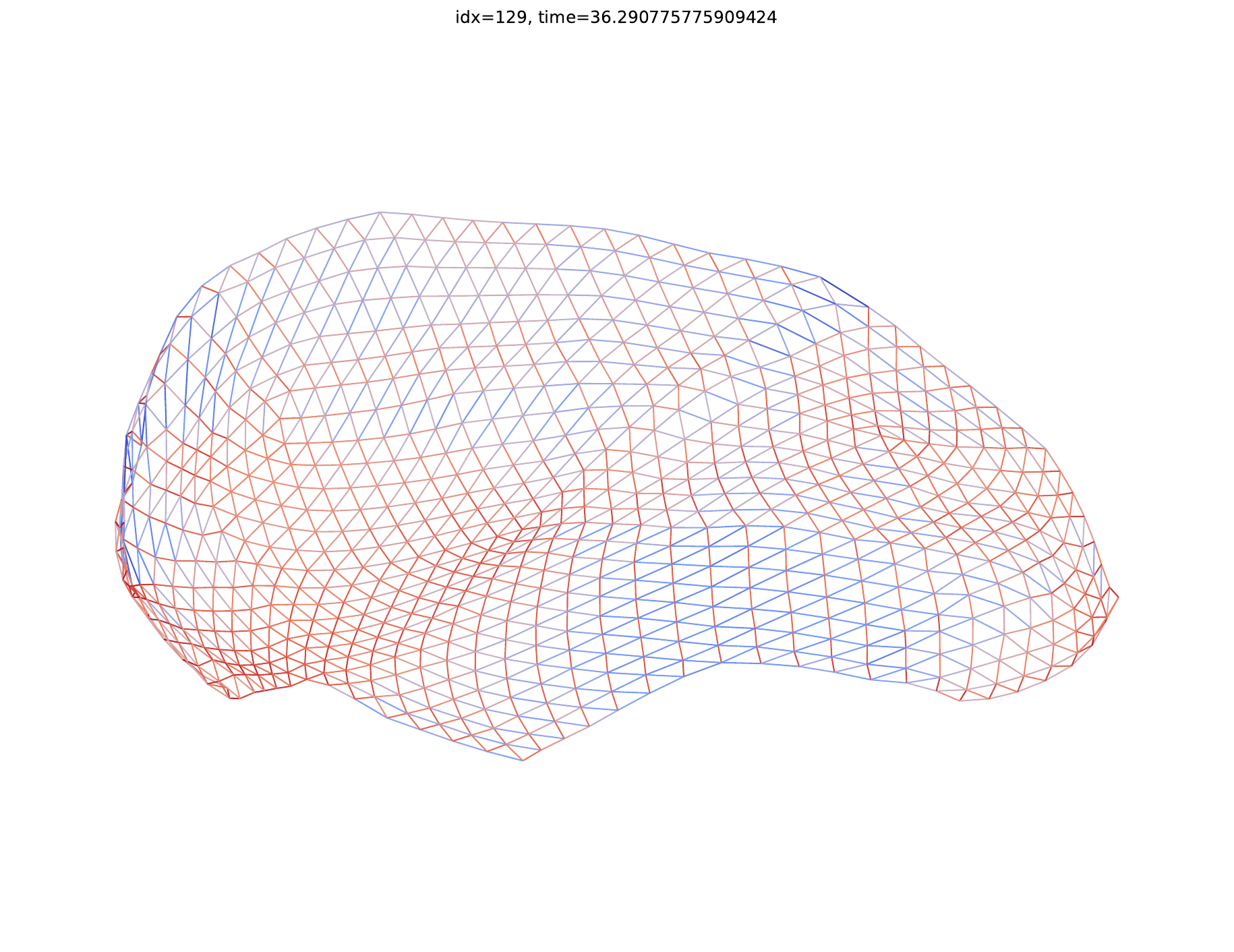} &
\imgcell{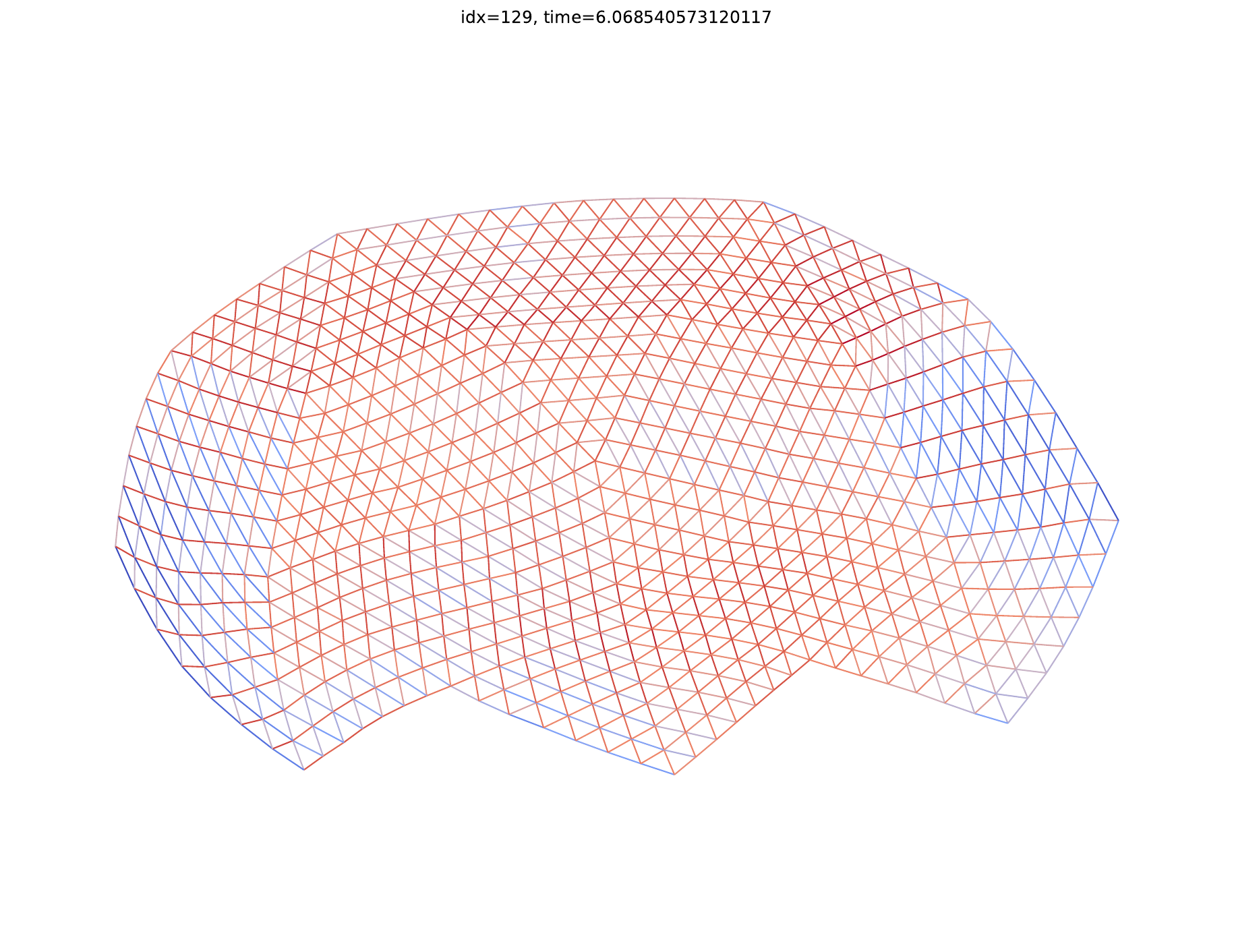} &
\imgcell{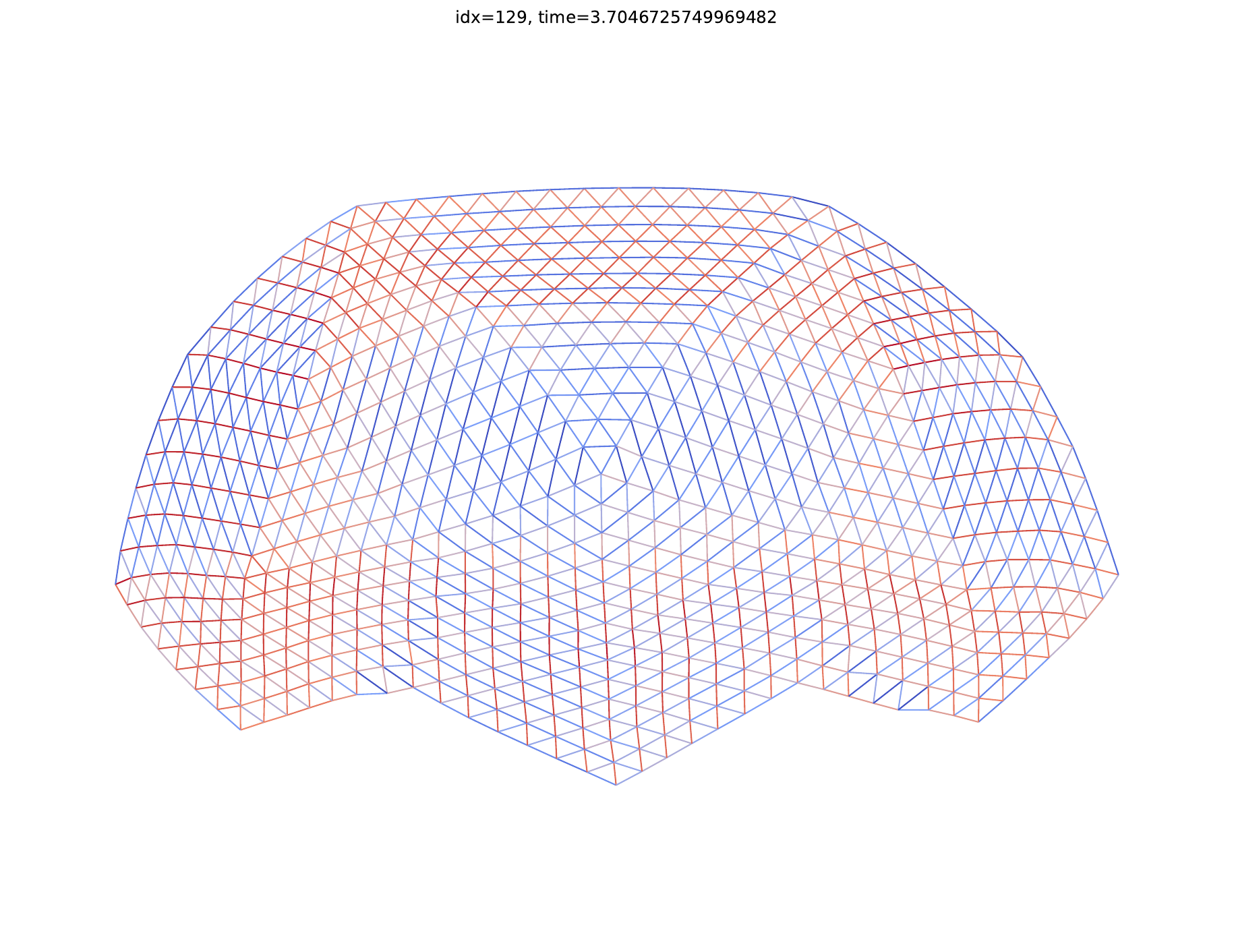} &
\imgcell{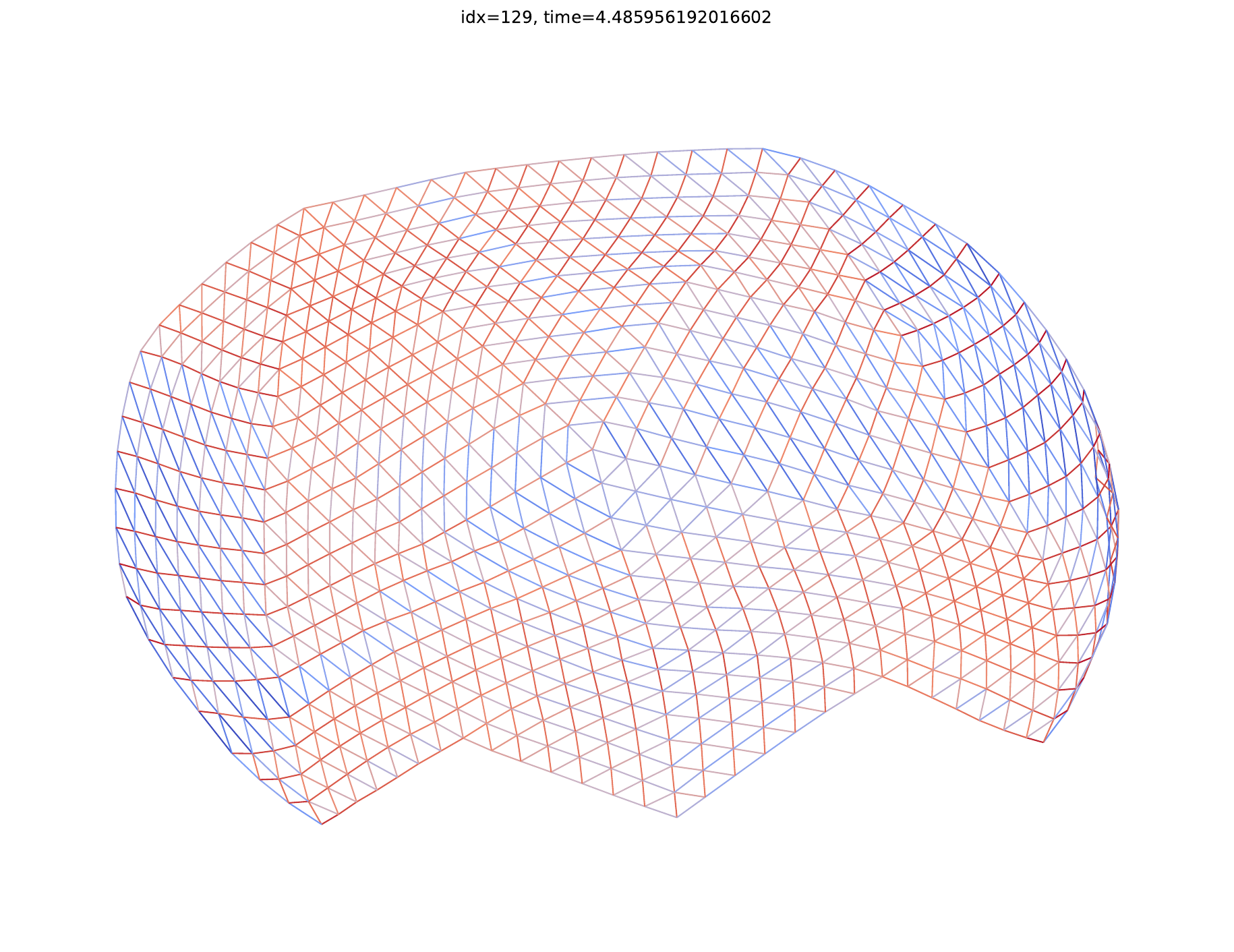} \\

&
t = 0.24s &
t = 8.08s &
t = 46.19s &
t = 2.42s &
t = 7200.00s &
t = 2.49s &
t = 2.46s &
t = 2.45s &
t = 2.69s &
t = 2.71s &
t = 2.86s &
t = 2.76s \\

\makecell{\bfseries bfwa782\\N = 782\\M = 3394} &
\imgcell{figures/large_graphs/130_sgd2.pdf} &
\imgcell{figures/large_graphs/130_pmds.pdf} &
\imgcell{figures/large_graphs/130_fa2.pdf} &
\imgcell{figures/large_graphs/130_deepgd.pdf} &
\imgcell{figures/large_graphs/130_gd2_stress_xing.pdf} &
\imgcell{figures/large_graphs/130_smartgd_stress.pdf} &
\imgcell{figures/large_graphs/130_smartgd_xing.pdf} &
\imgcell{figures/large_graphs/130_smartgd_xing_nsc.pdf} &
\imgcell{figures/large_graphs/130_smartgd_xangle.pdf} &
\imgcell{figures/large_graphs/130_smartgd_stress_xing.pdf} &
\imgcell{figures/large_graphs/130_smartgd_stress_xangle.pdf} &
\imgcell{figures/large_graphs/130_smartgd_combined.pdf} \\

&
t = 0.21s &
t = 20.45s &
t = 48.10s &
t = 2.52s &
t = 7200.00s &
t = 2.84s &
t = 2.60s &
t = 2.93s &
t = 2.60s &
t = 2.83s &
t = 2.68s &
t = 2.67s \\

\makecell{\bfseries can\_838\\N = 838\\M = 4586} &
\imgcell{figures/large_graphs/164_sgd2.pdf} &
\imgcell{figures/large_graphs/164_pmds.pdf} &
\imgcell{figures/large_graphs/164_fa2.pdf} &
\imgcell{figures/large_graphs/164_deepgd.pdf} &
\imgcell{figures/large_graphs/164_gd2_stress_xing.pdf} &
\imgcell{figures/large_graphs/164_smartgd_stress.pdf} &
\imgcell{figures/large_graphs/164_smartgd_xing.pdf} &
\imgcell{figures/large_graphs/164_smartgd_xing_nsc.pdf} &
\imgcell{figures/large_graphs/164_smartgd_xangle.pdf} &
\imgcell{figures/large_graphs/164_smartgd_stress_xing.pdf} &
\imgcell{figures/large_graphs/164_smartgd_stress_xangle.pdf} &
\imgcell{figures/large_graphs/164_smartgd_combined.pdf} \\

&
t = 0.25s &
t = 20.61s &
t = 60.44s &
t = 2.81s &
t = 7200.00s &
t = 3.27s &
t = 3.11s &
t = 3.30s &
t = 3.01s &
t = 3.35s &
t = 2.80s &
t = 2.61s \\

\makecell{\bfseries ex27\\N = 974\\M = 19904} &
\imgcell{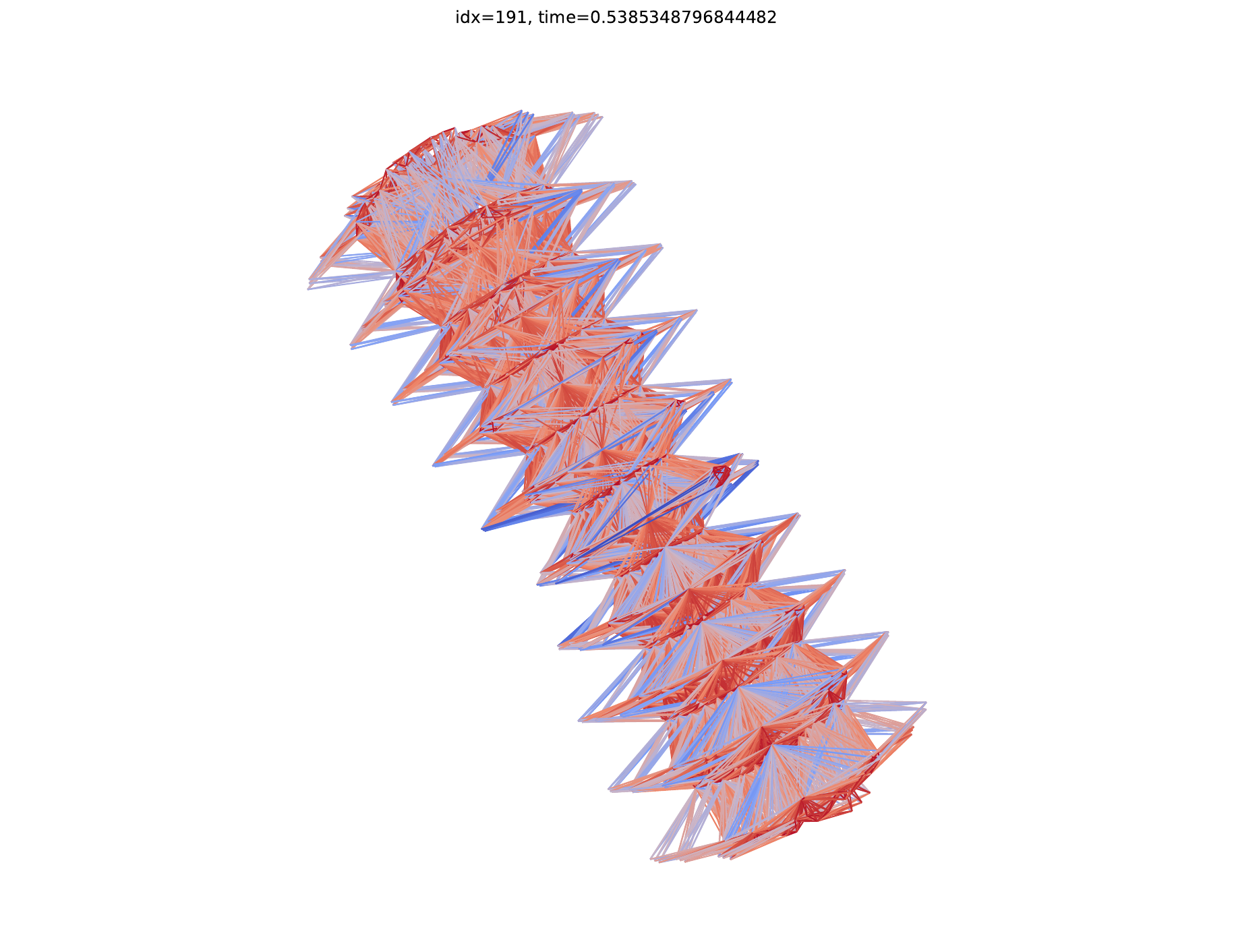} &
\imgcell{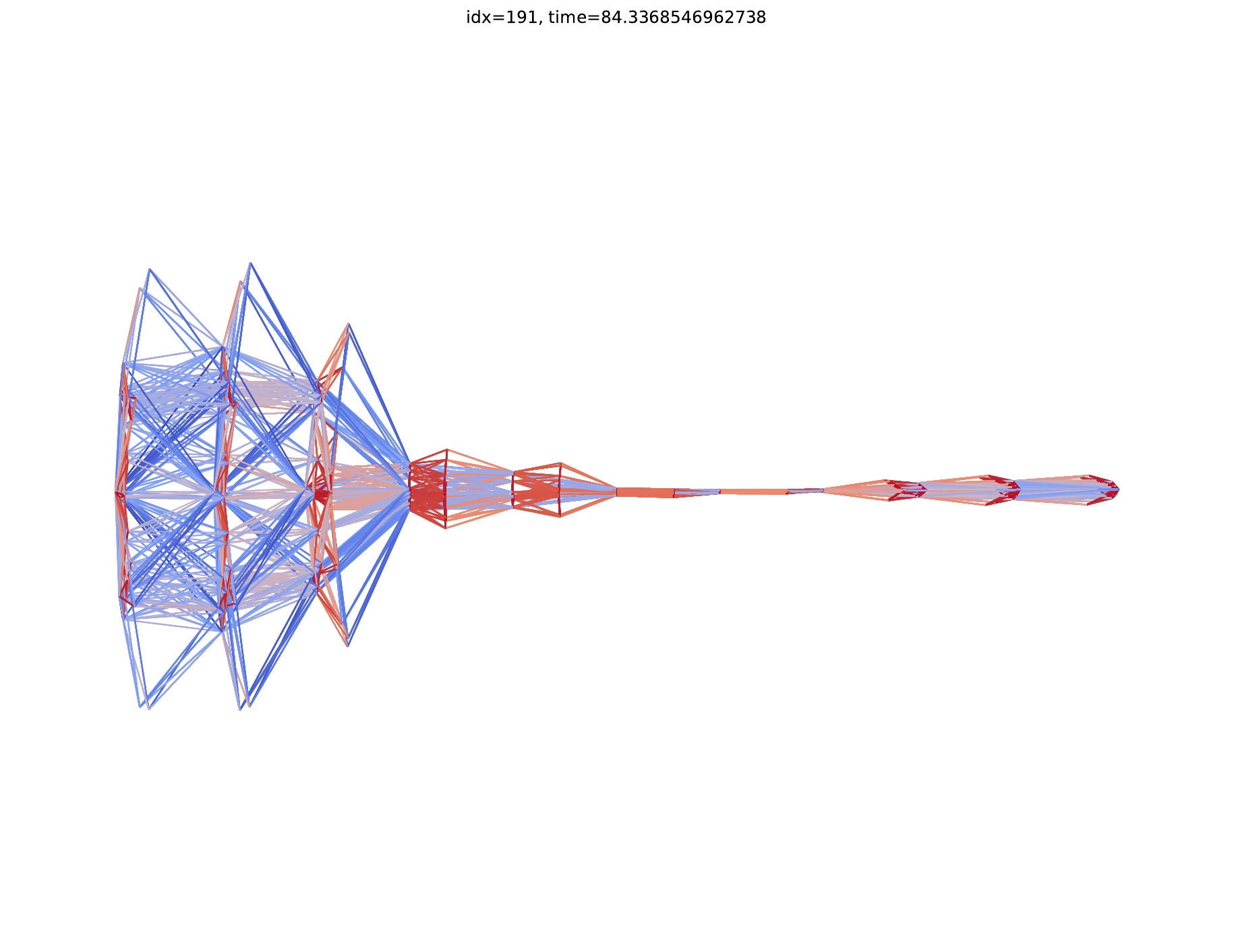} &
\imgcell{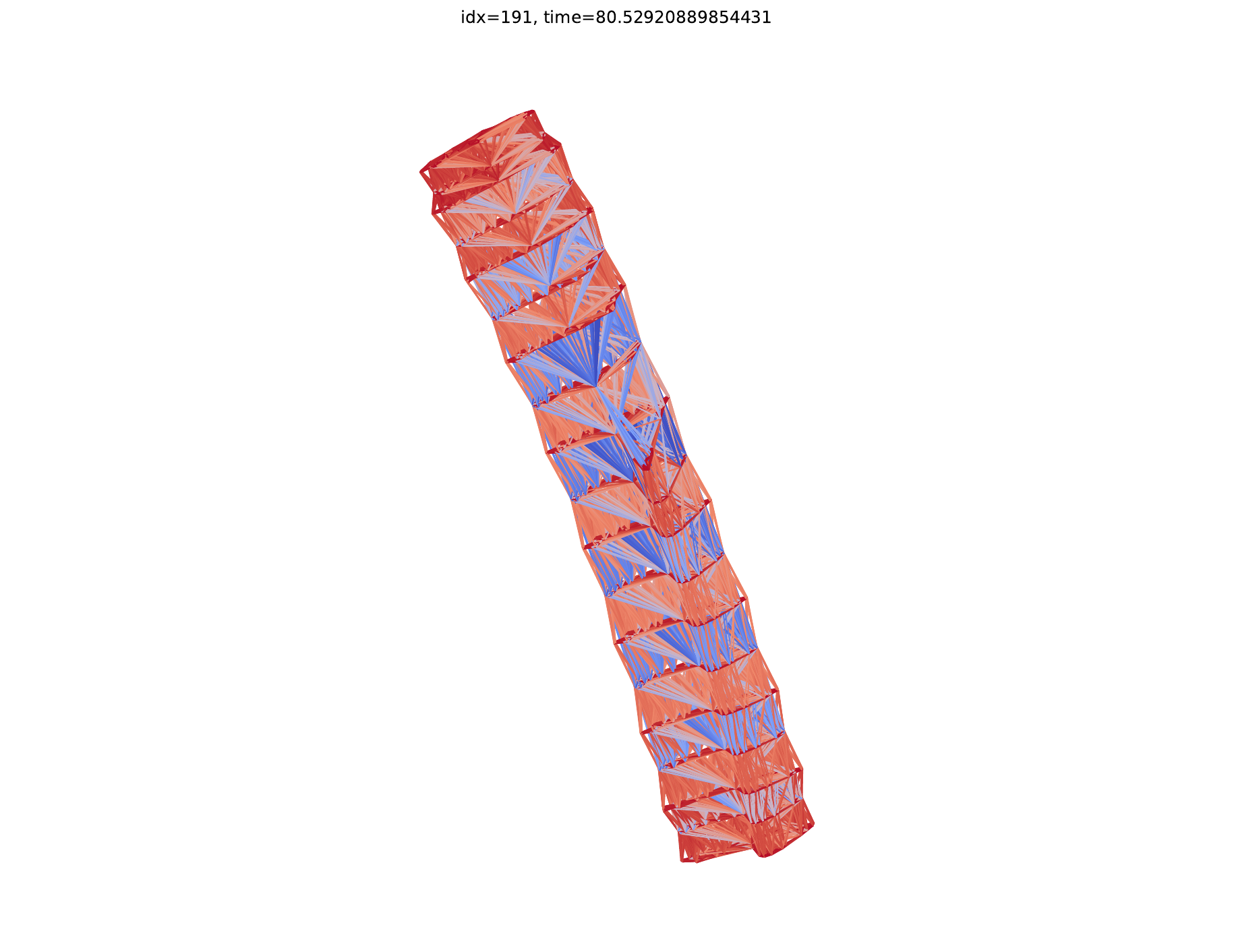} &
\imgcell{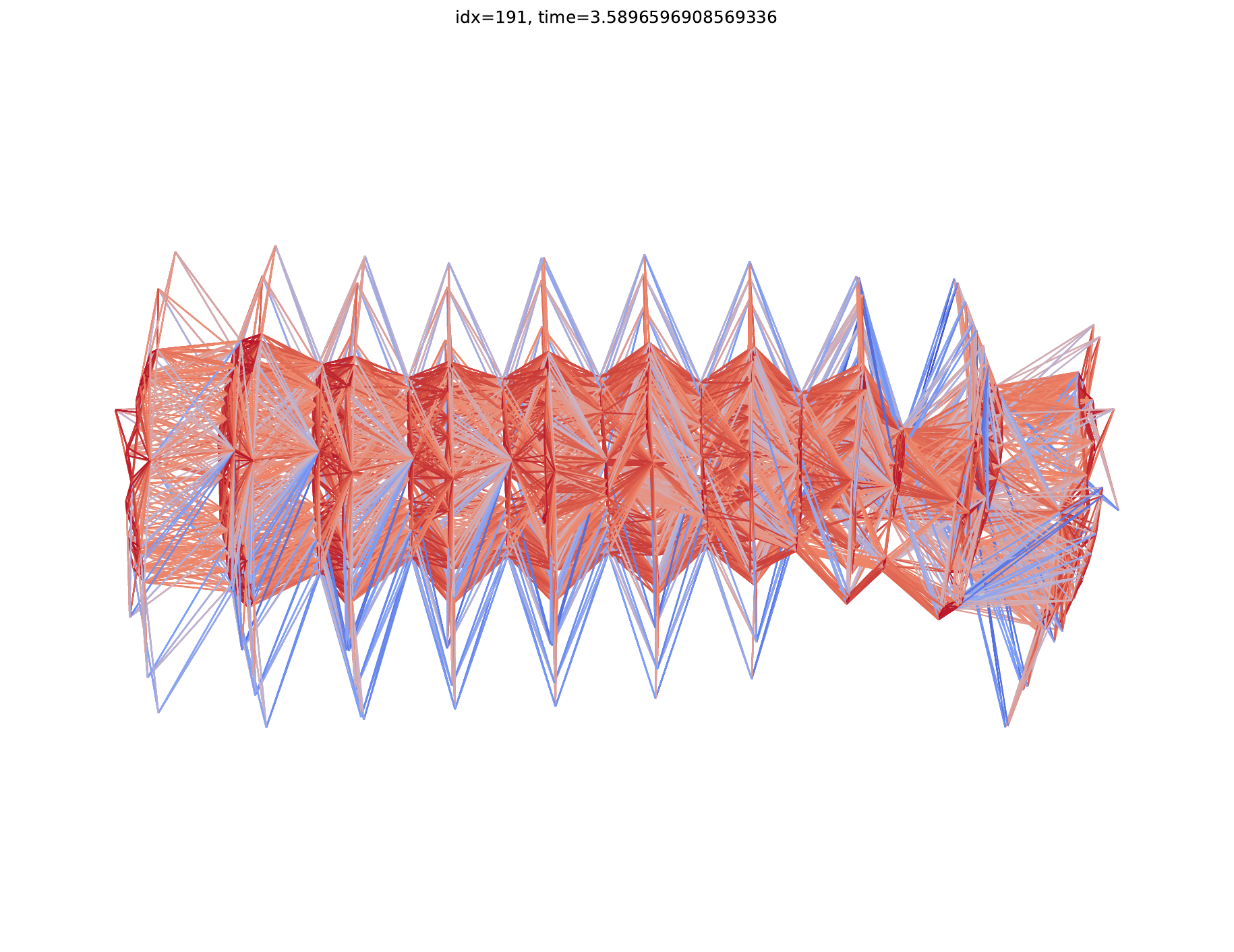} &
\imgcell{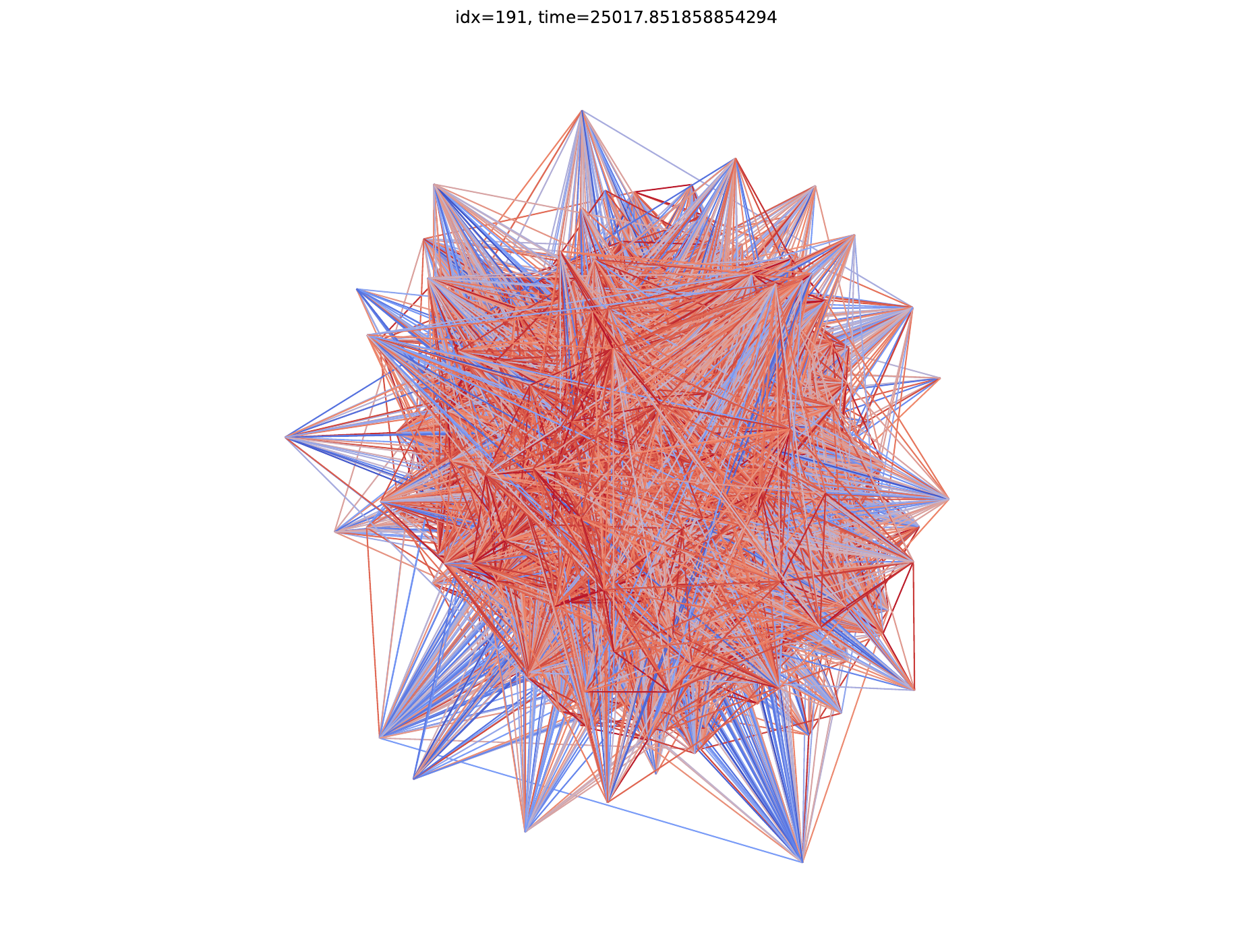} &
\imgcell{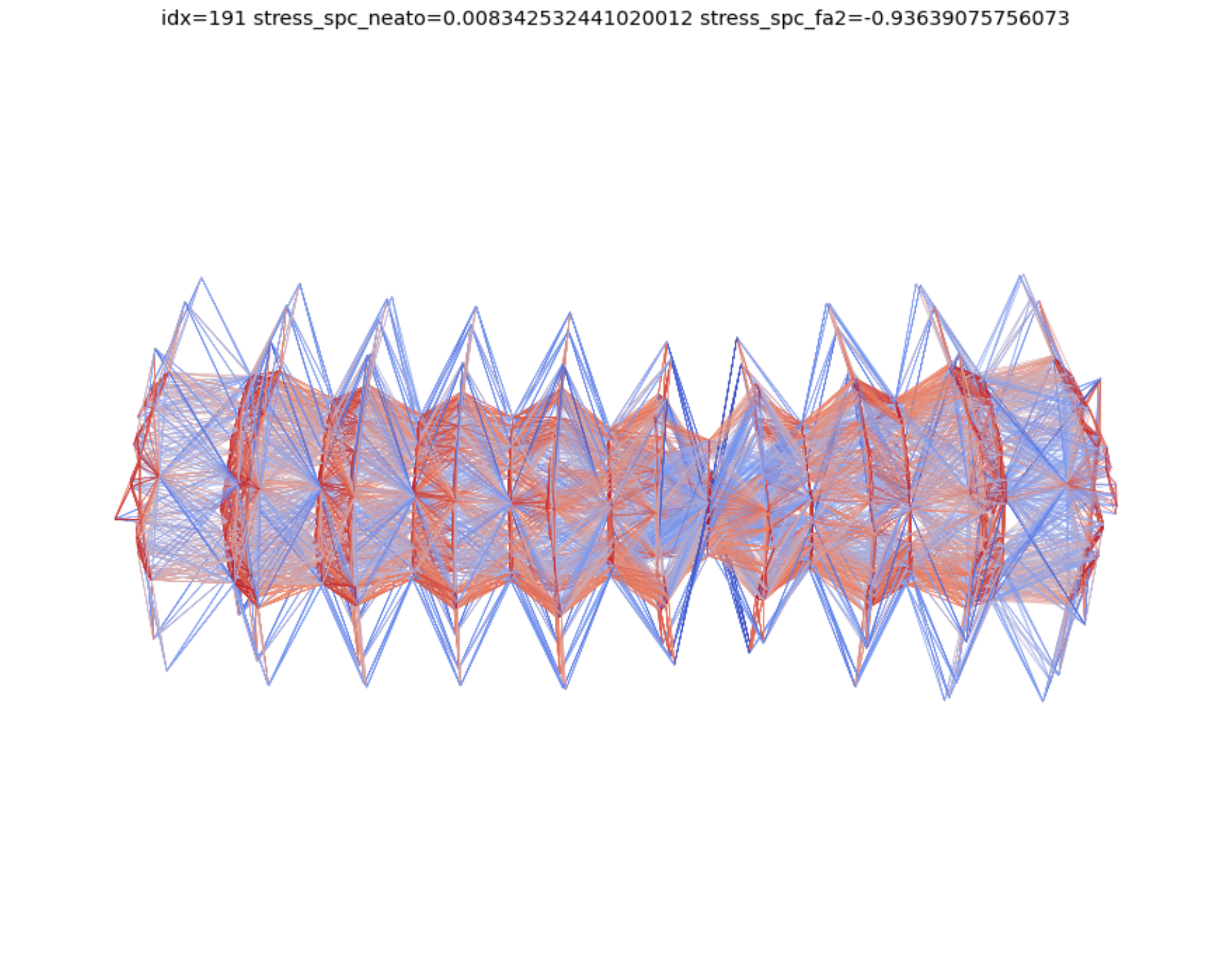} &
\imgcell{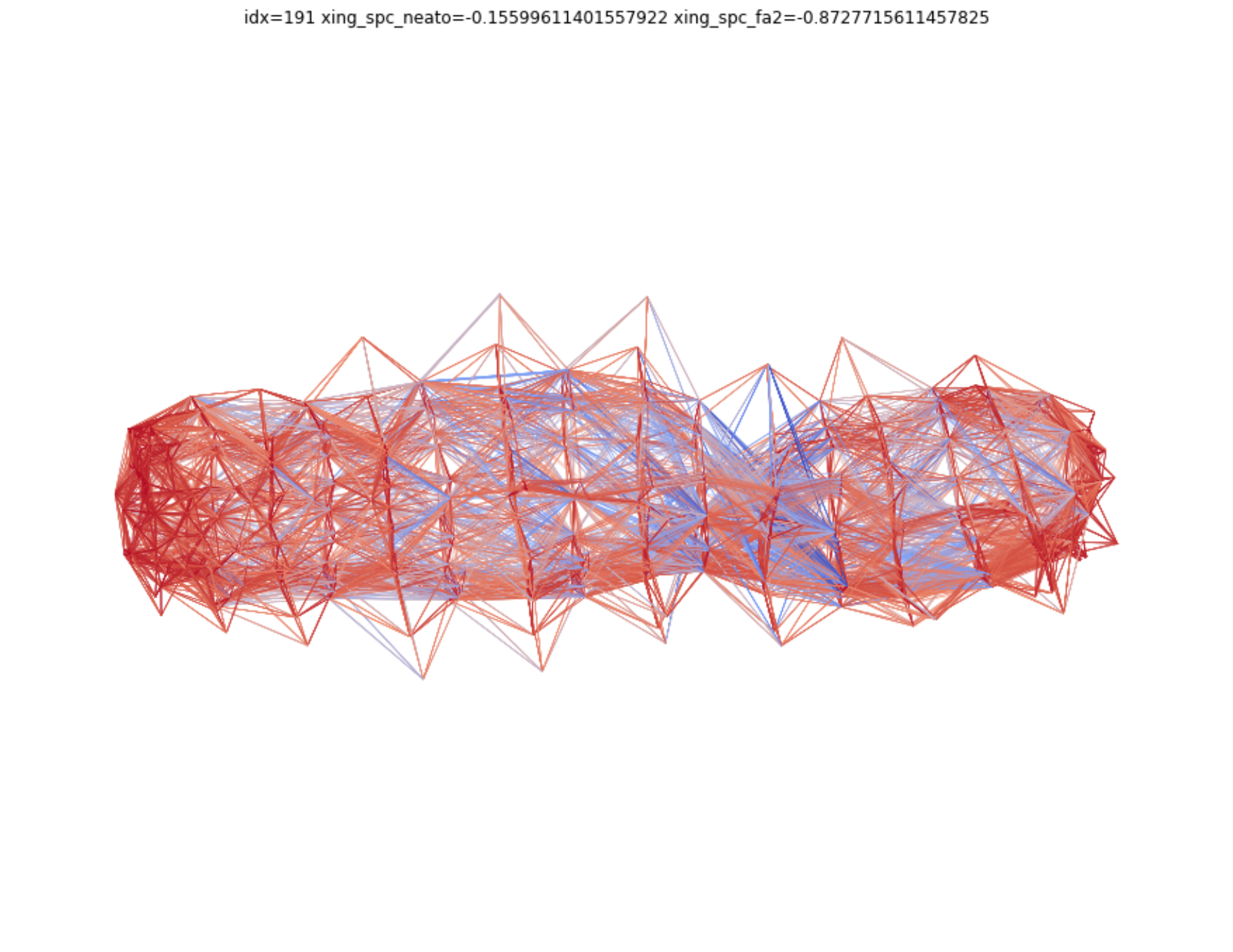} &
\imgcell{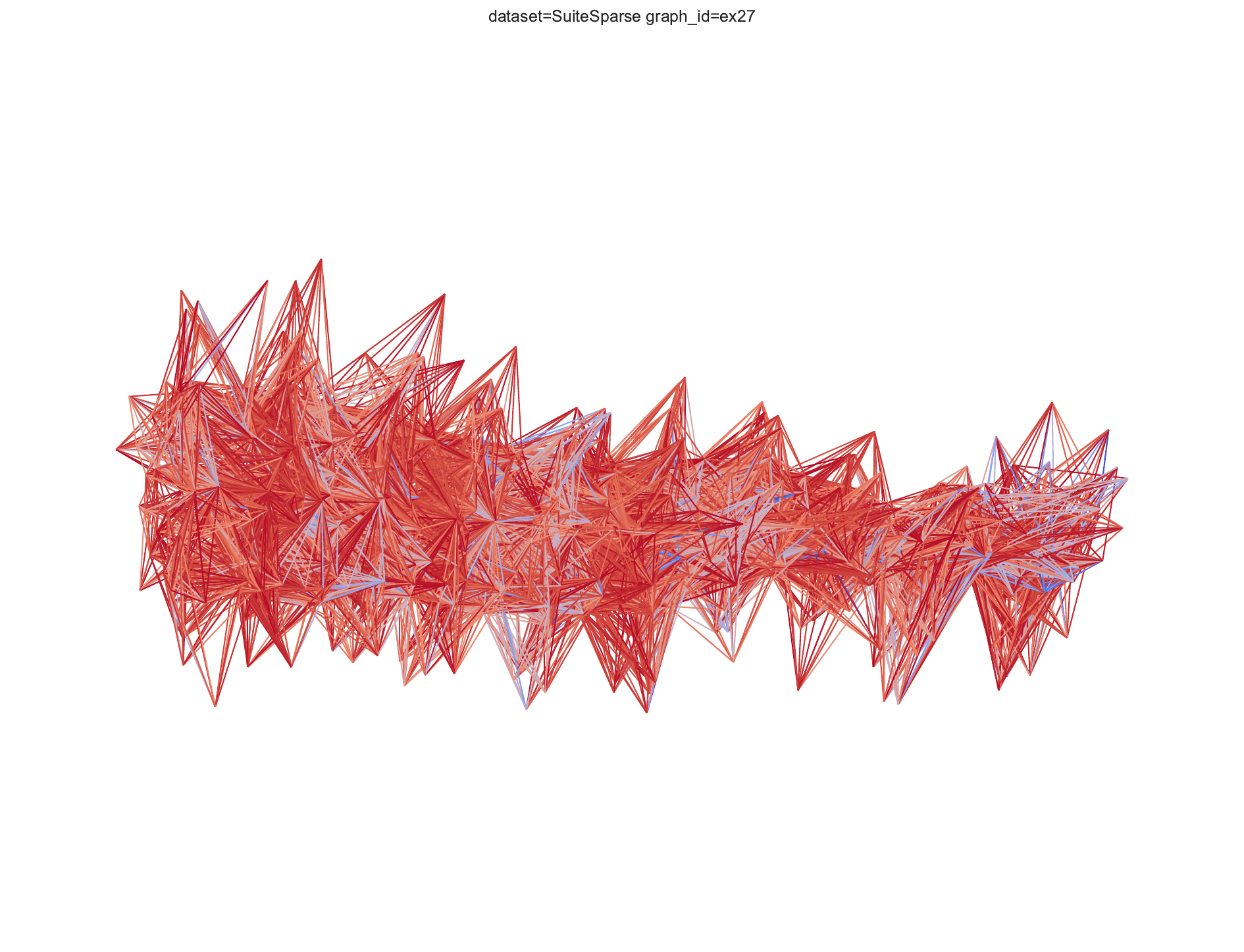} &
\imgcell{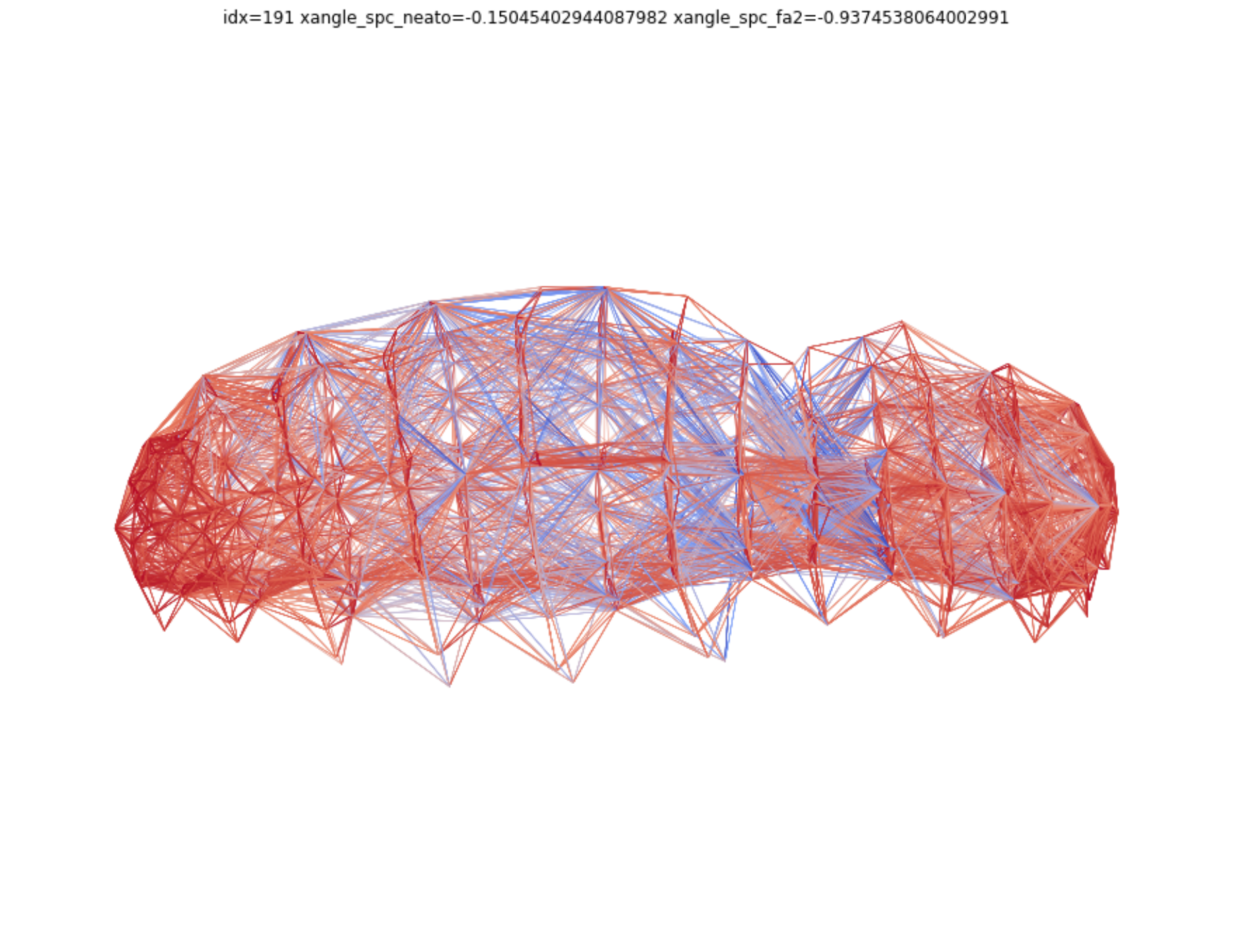} &
\imgcell{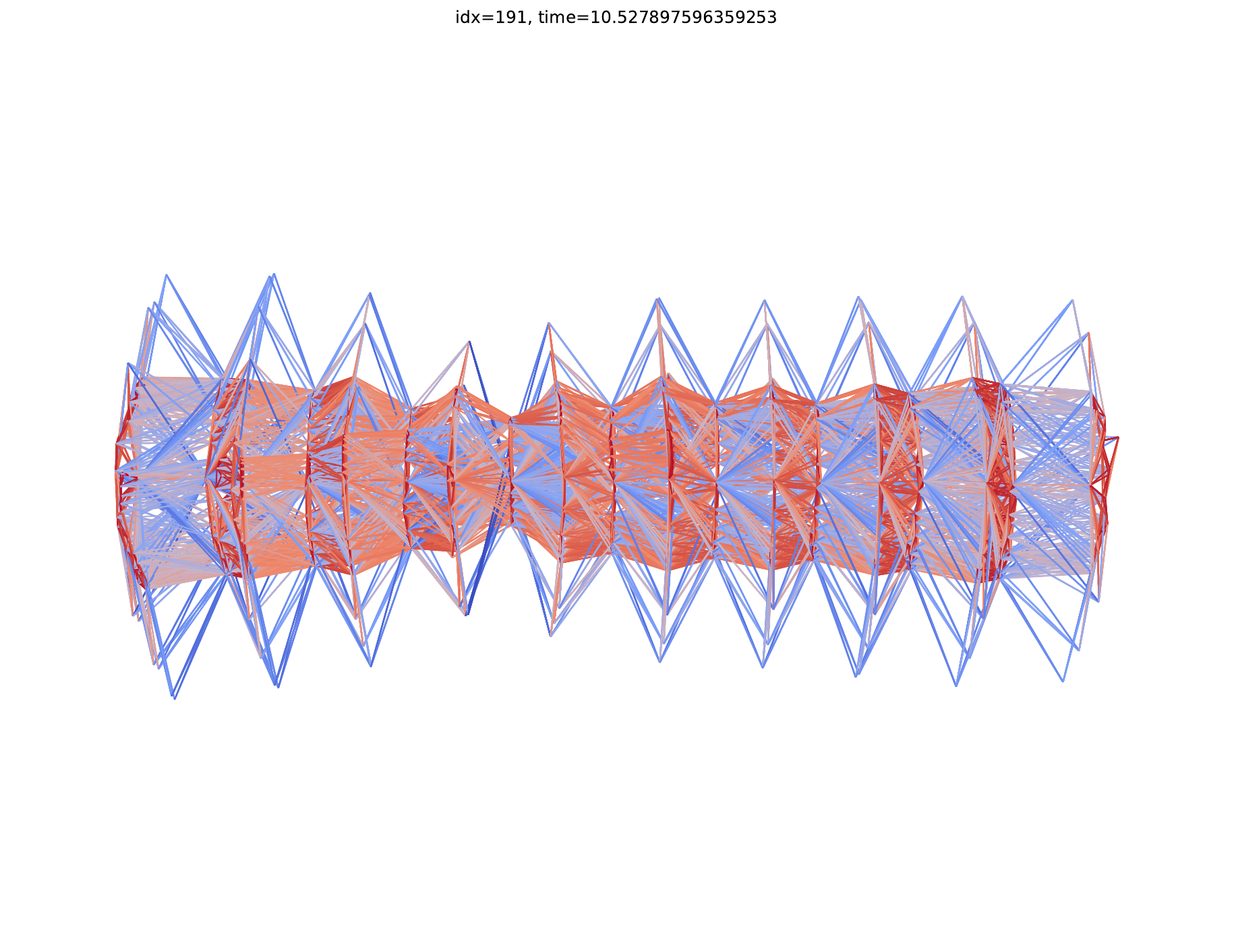} &
\imgcell{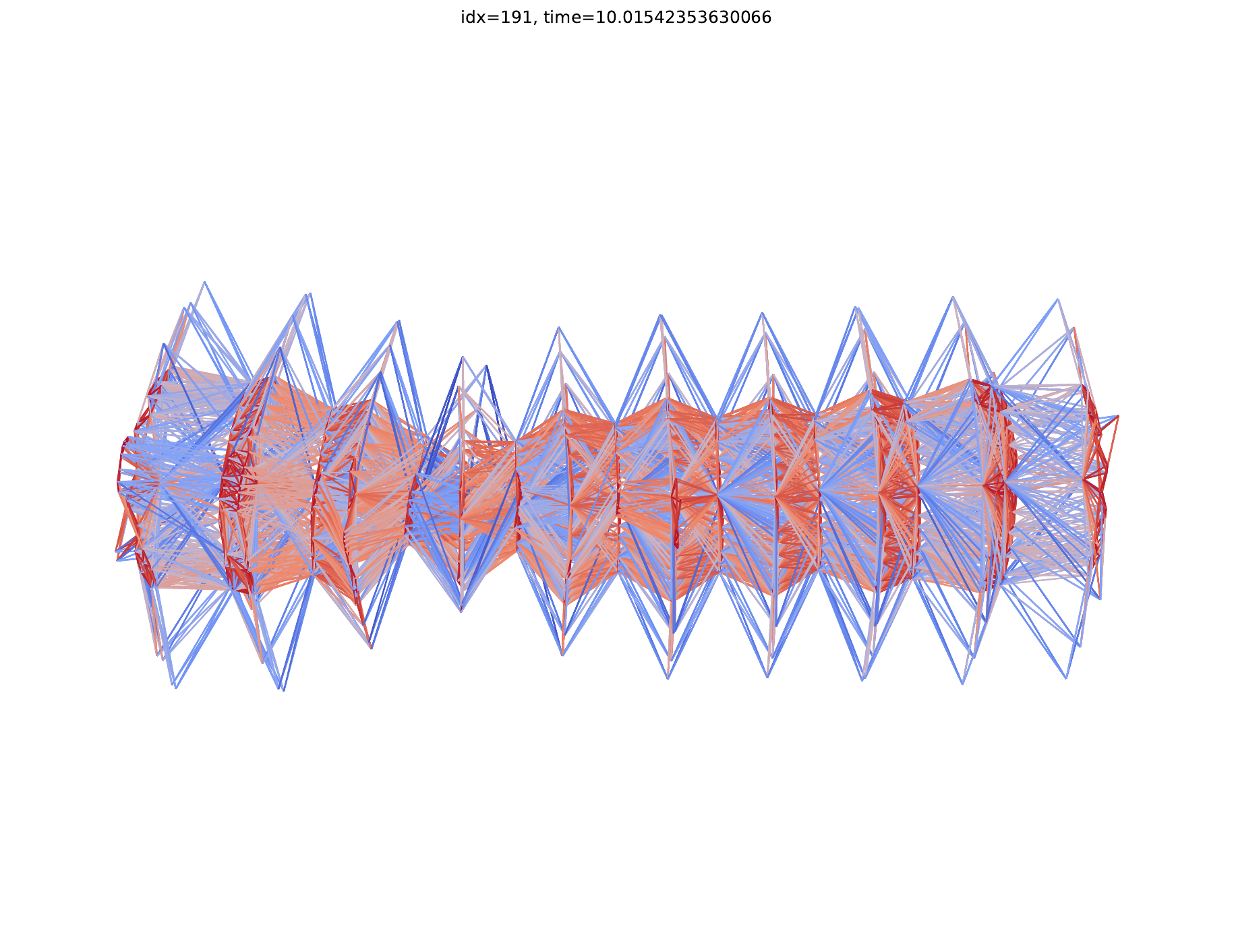} &
\imgcell{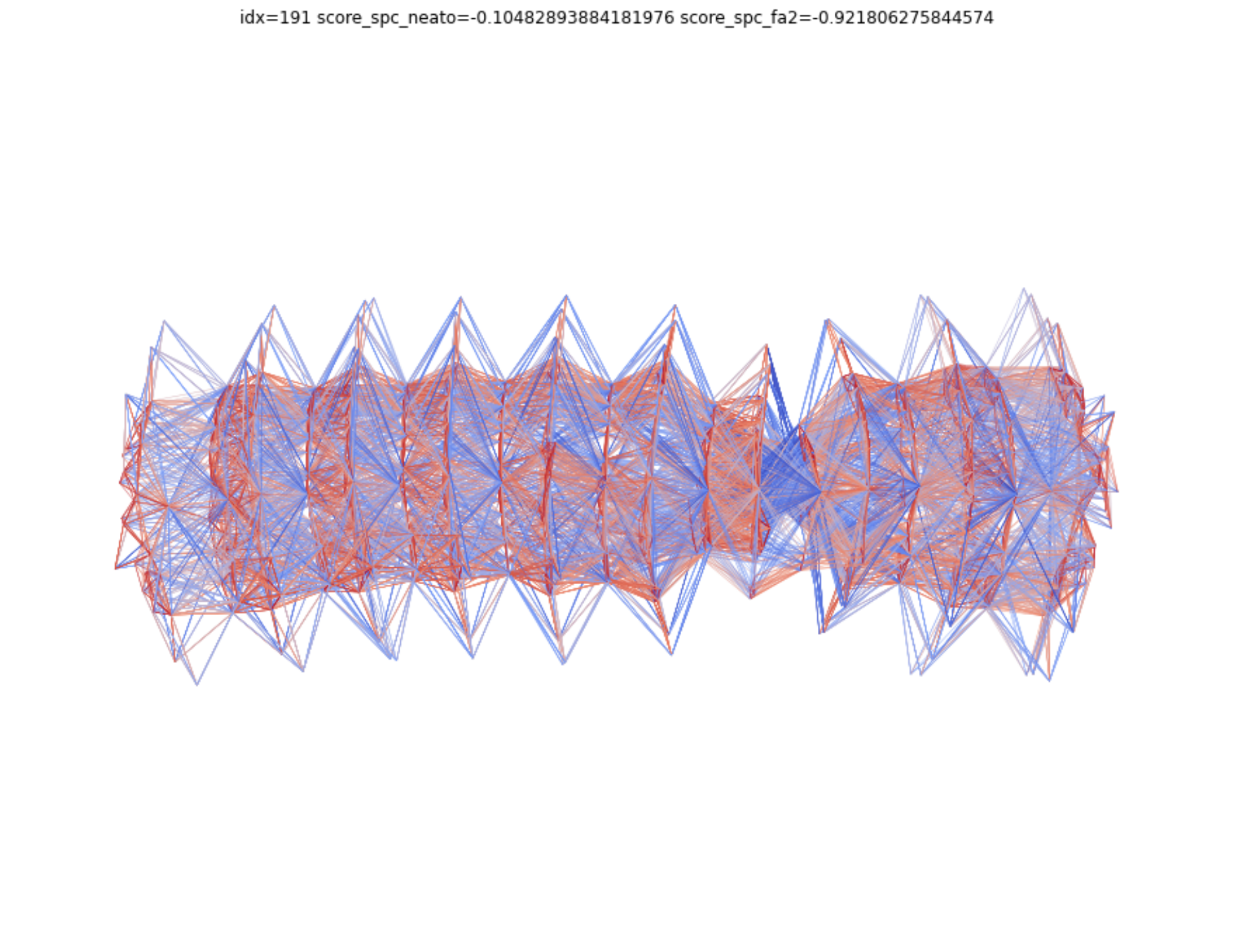} \\

&
t = 0.54s &
t = 84.34s &
t = 80.53s &
t = 3.59s &
t = 7200.00s &
t = 4.40s &
t = 3.26s &
t = 3.66s &
t = 3.84s &
t = 3.03s &
t = 4.02s &
t = 3.60s \\

\makecell{\bfseries circuit204\\N = 1020\\M = 3973} &
\imgcell{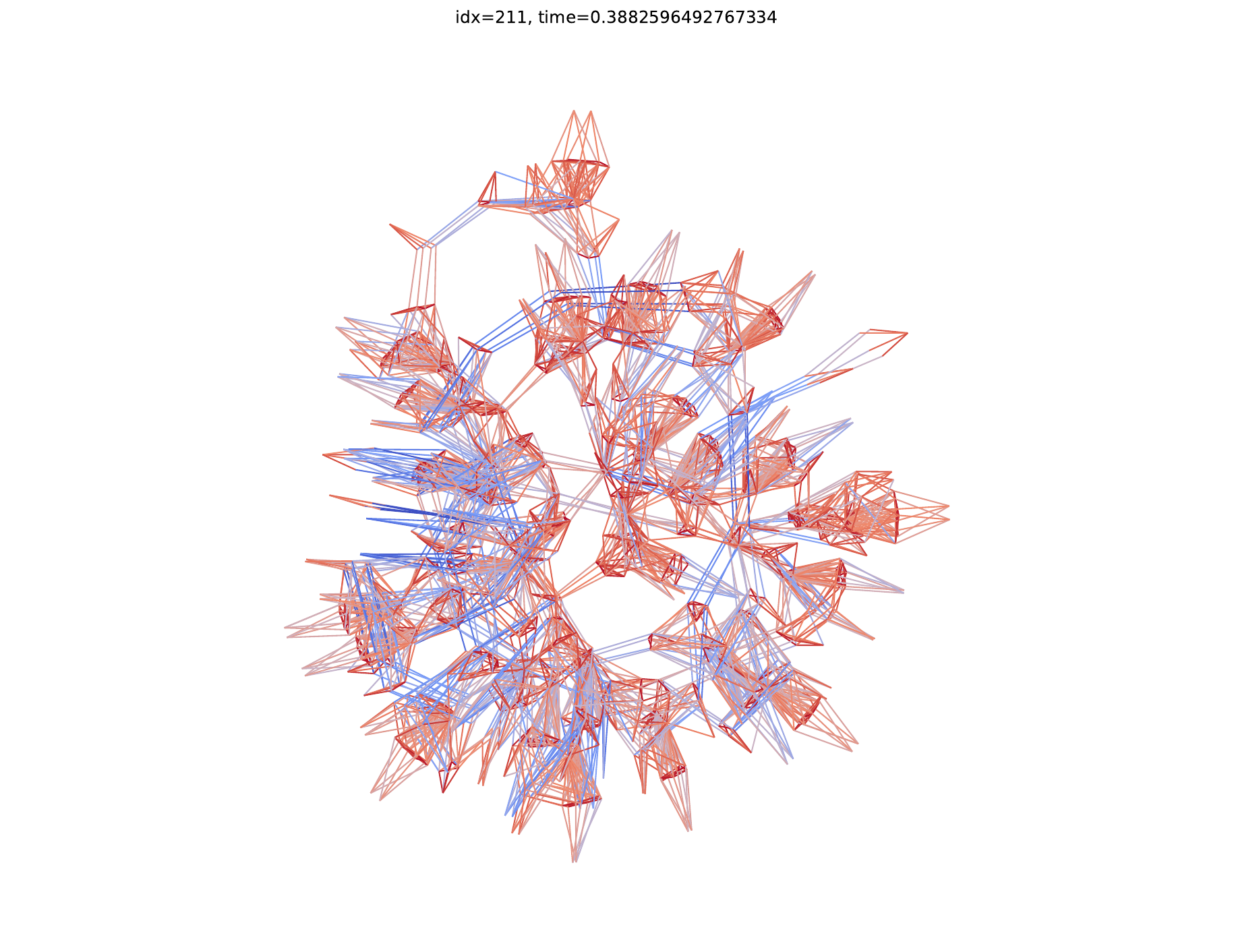} &
\imgcell{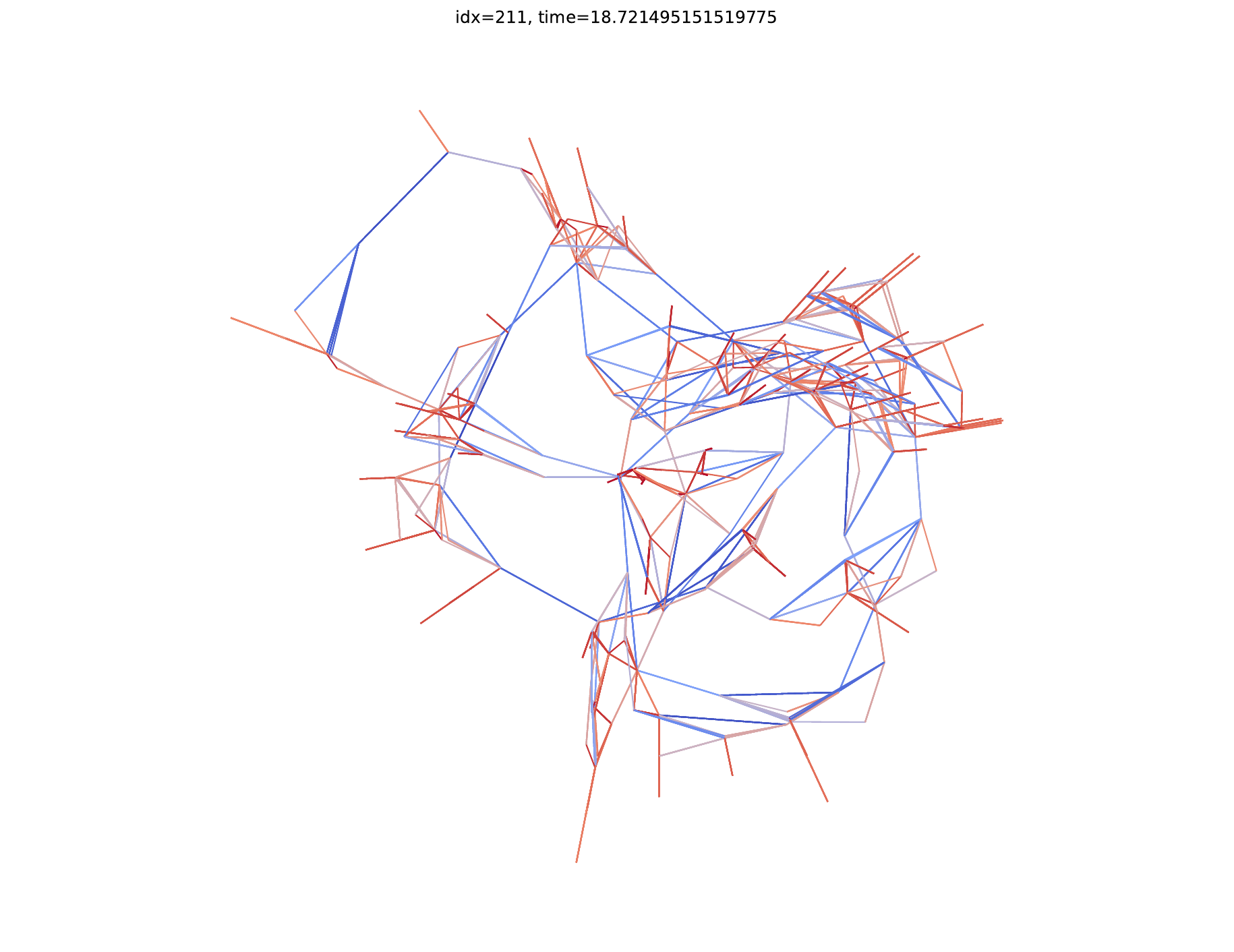} &
\imgcell{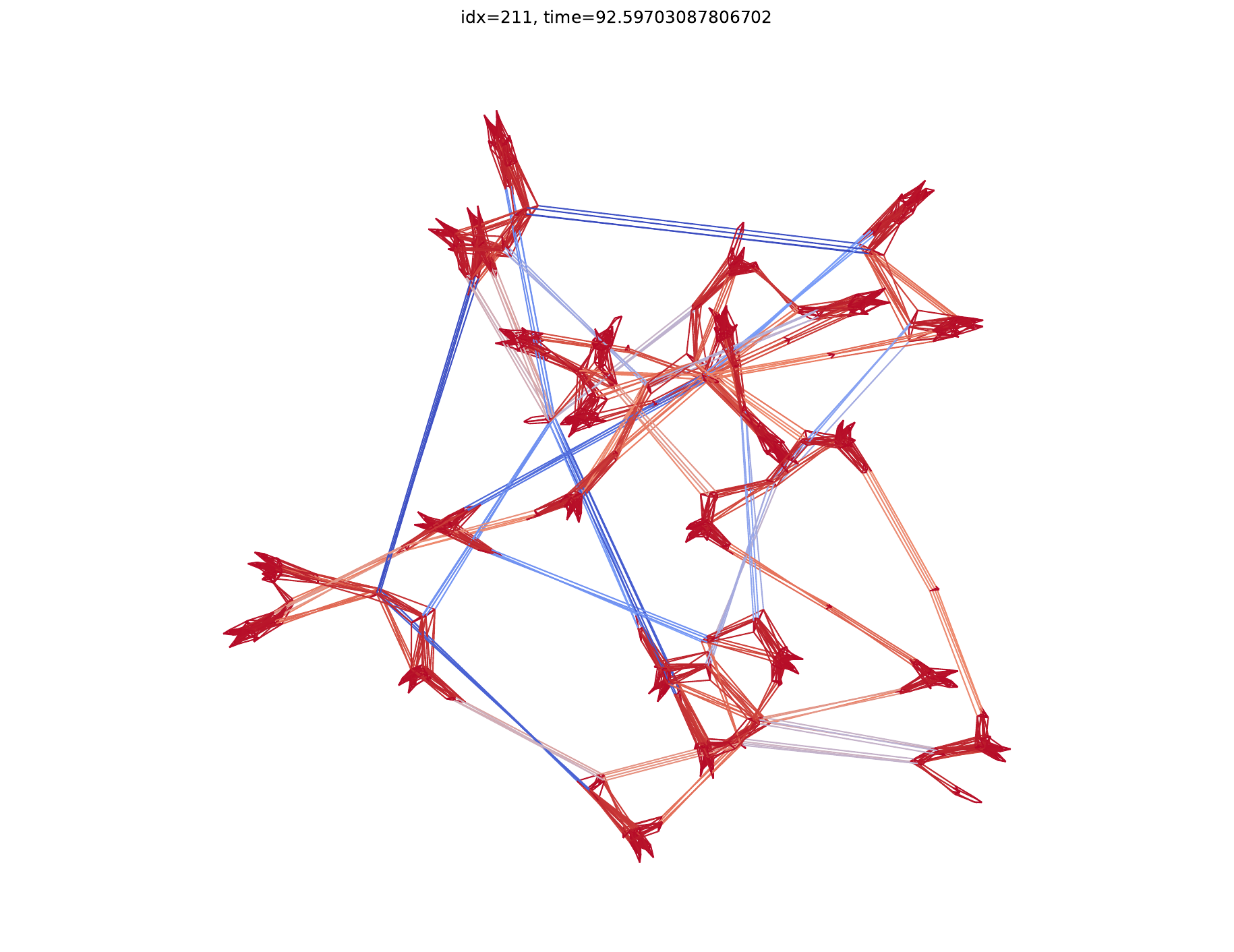} &
\imgcell{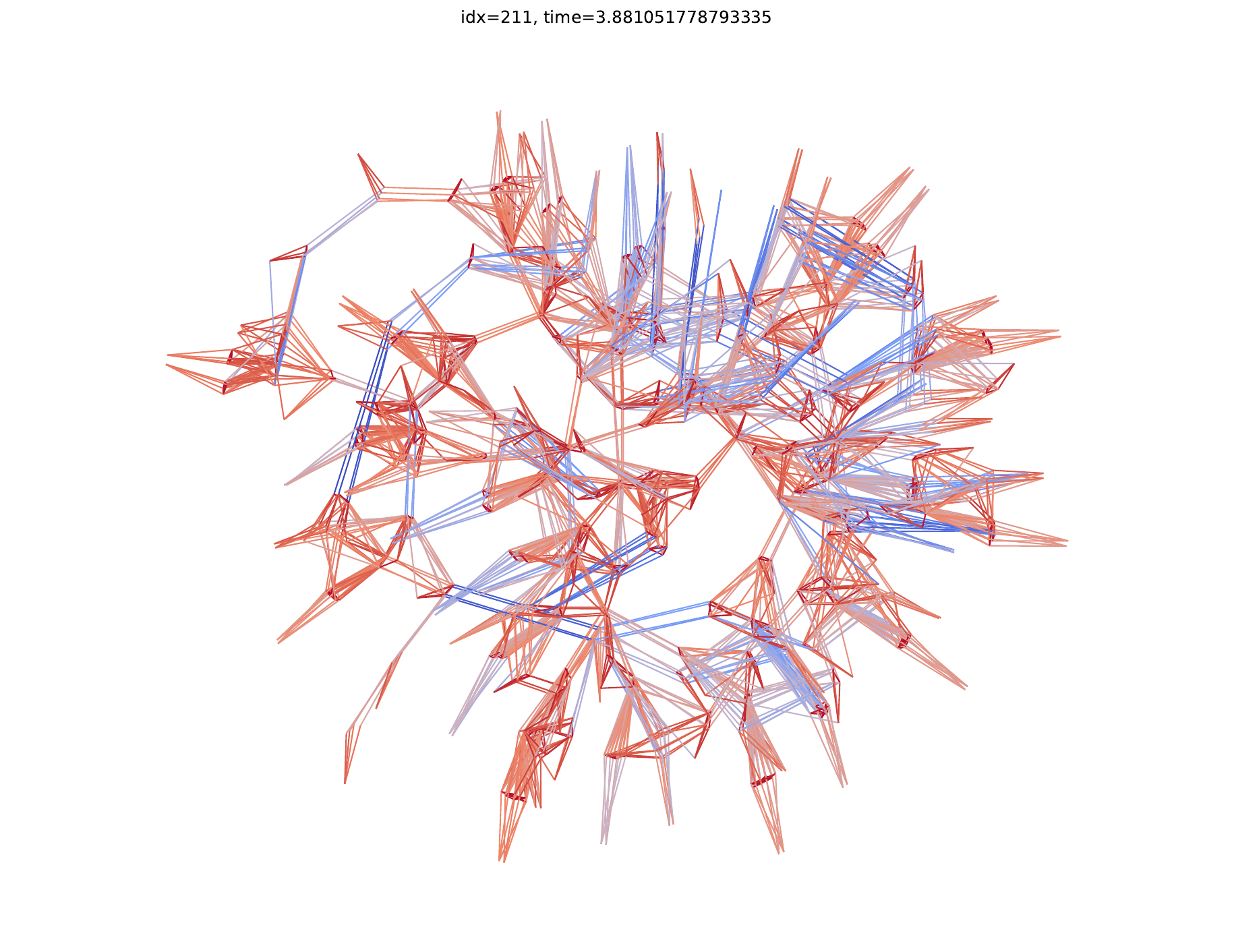} &
\imgcell{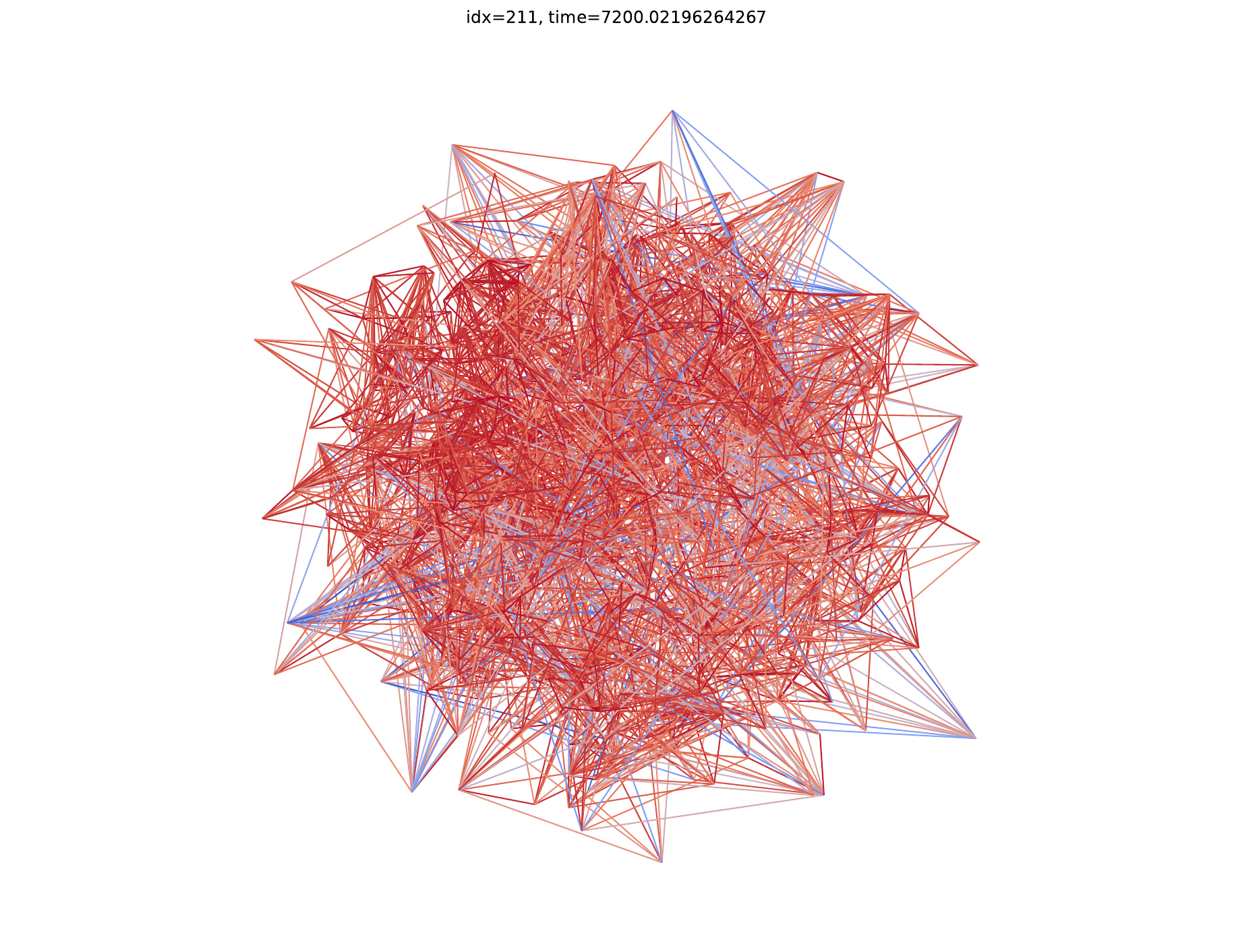} &
\imgcell{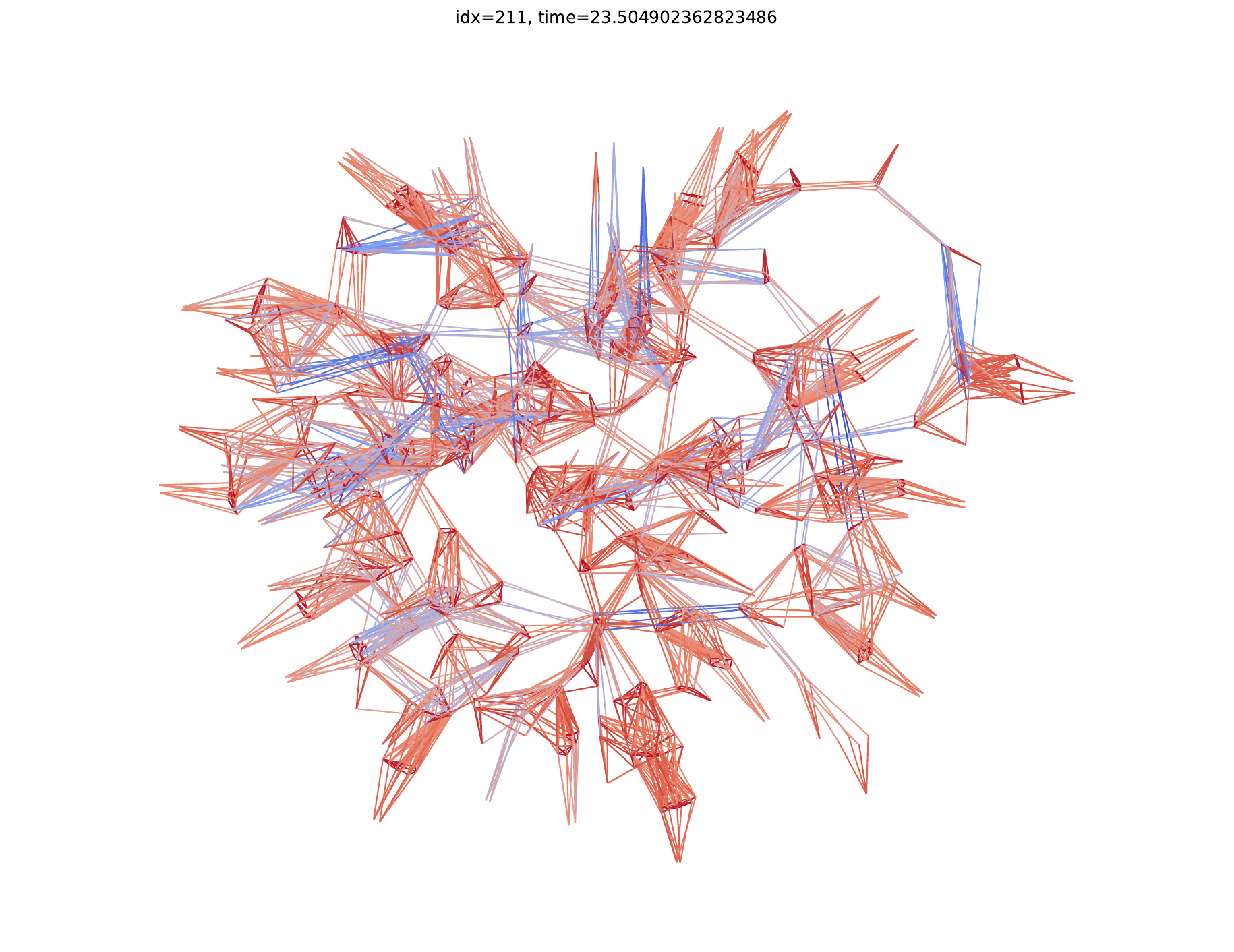} &
\imgcell{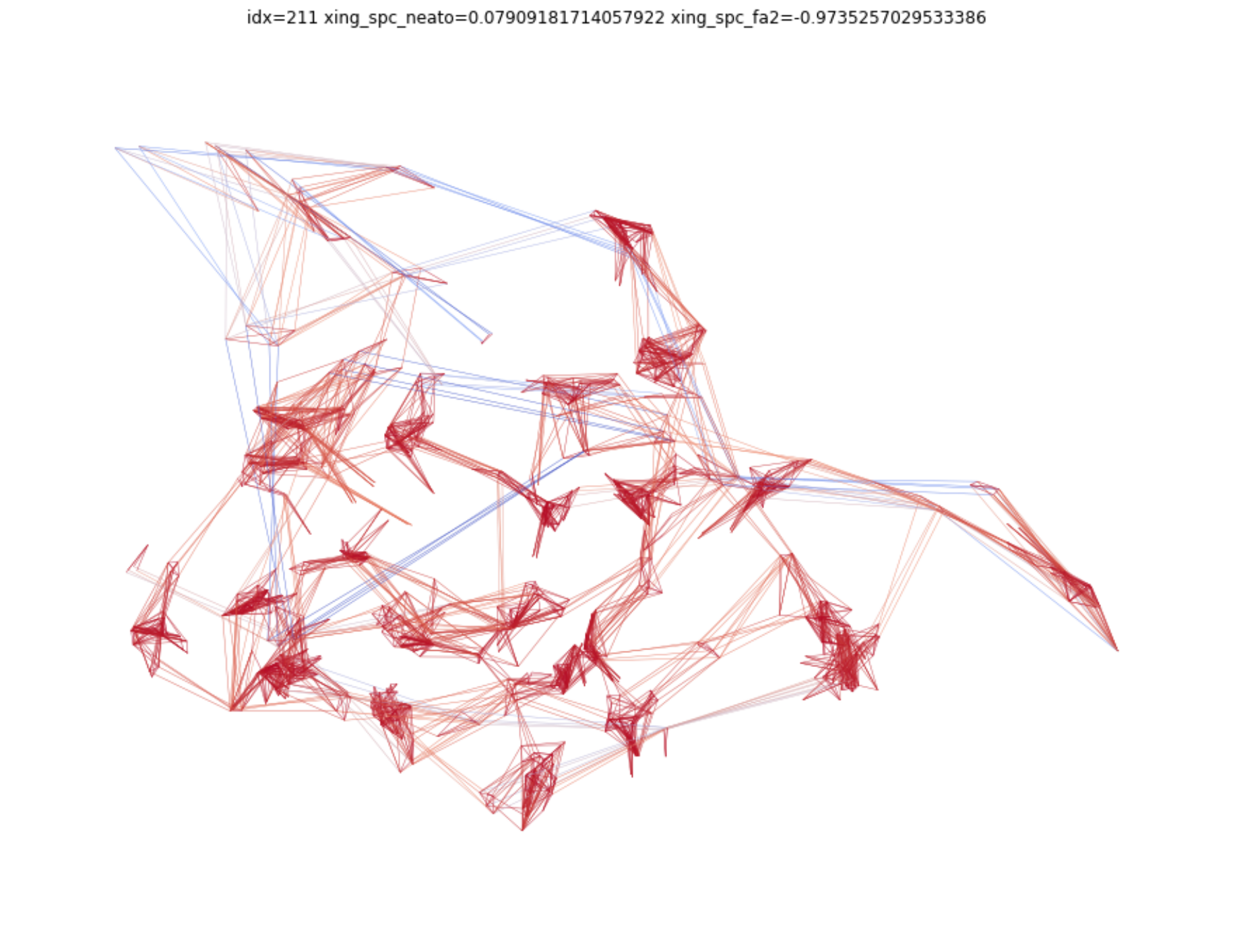} &
\imgcell{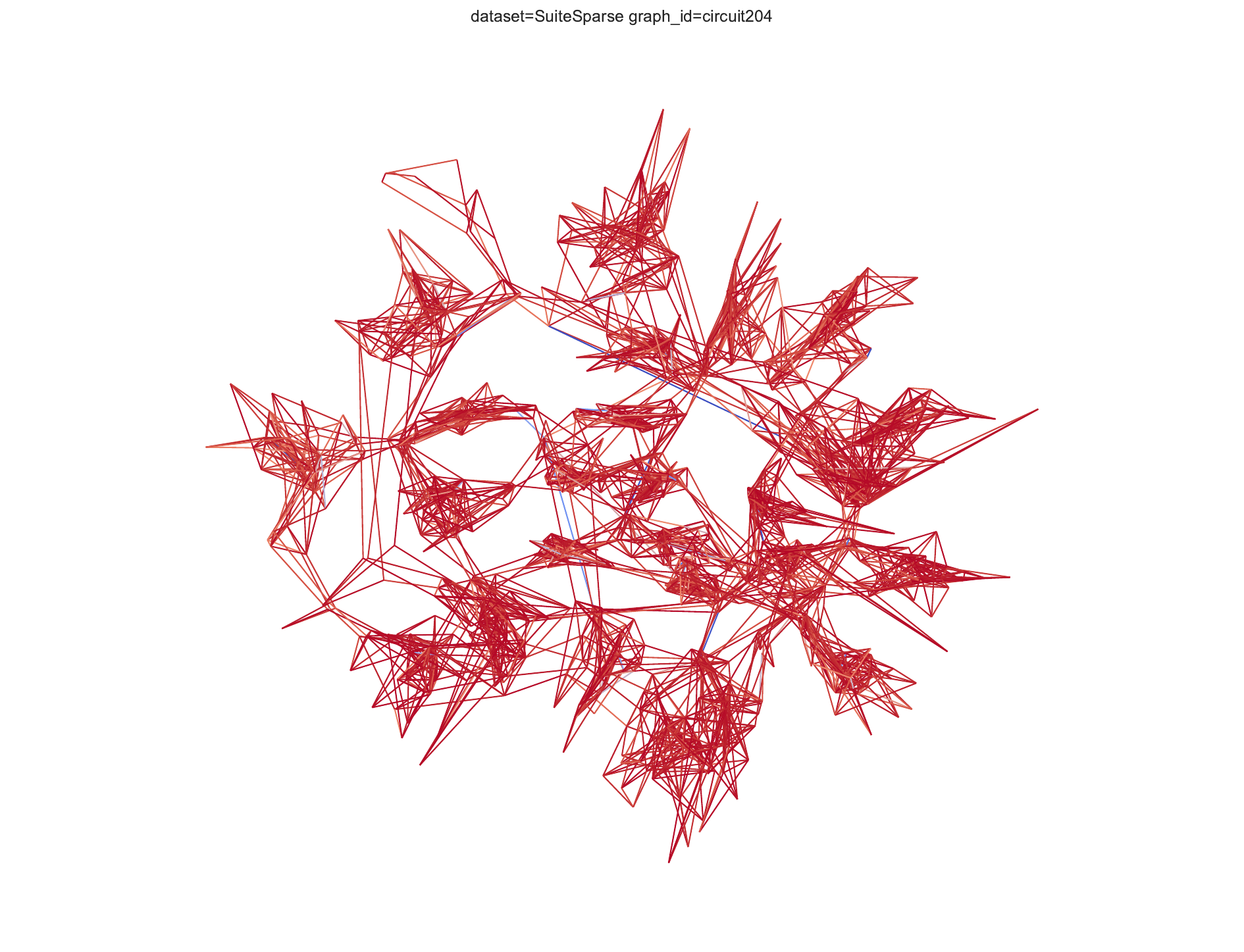} &
\imgcell{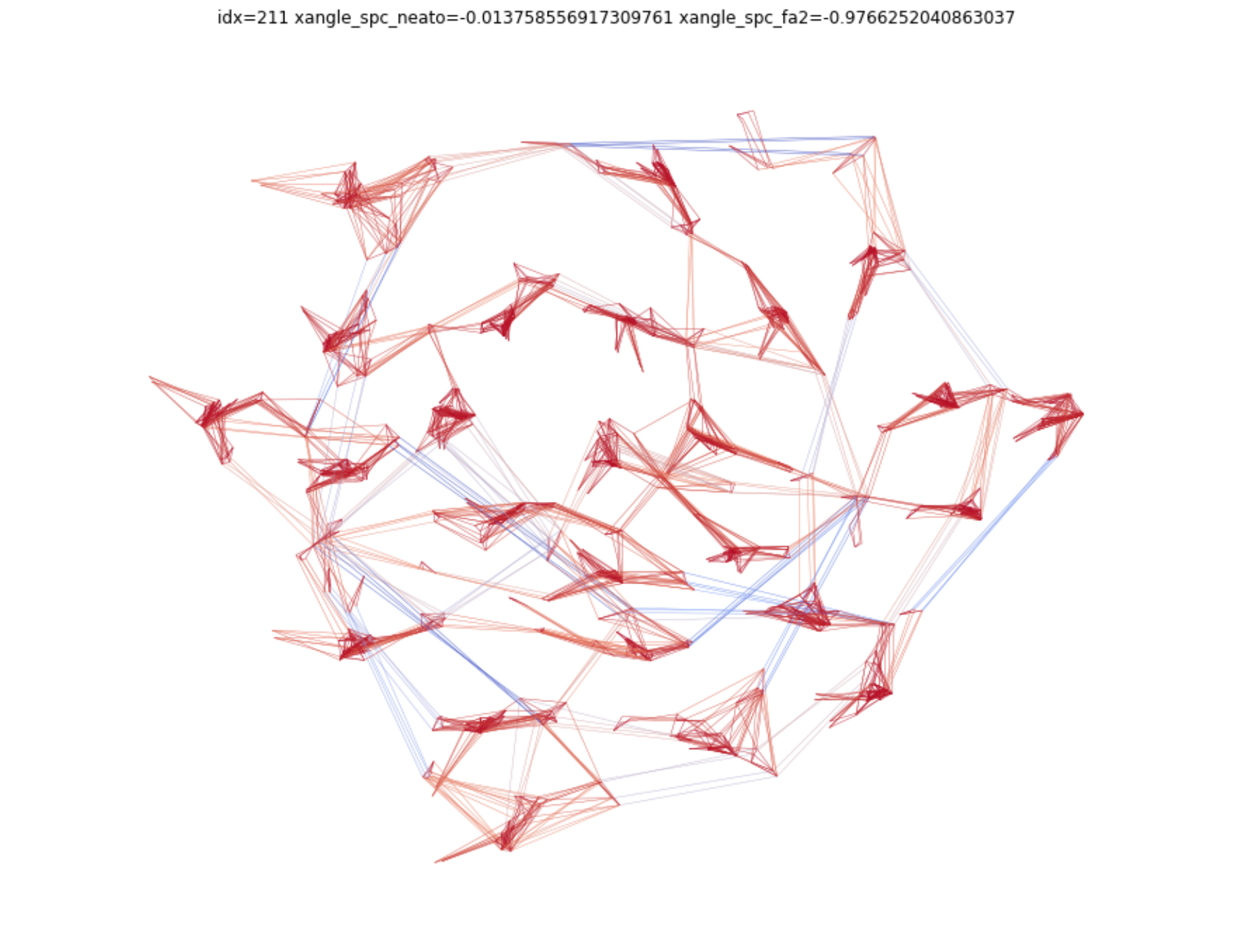} &
\imgcell{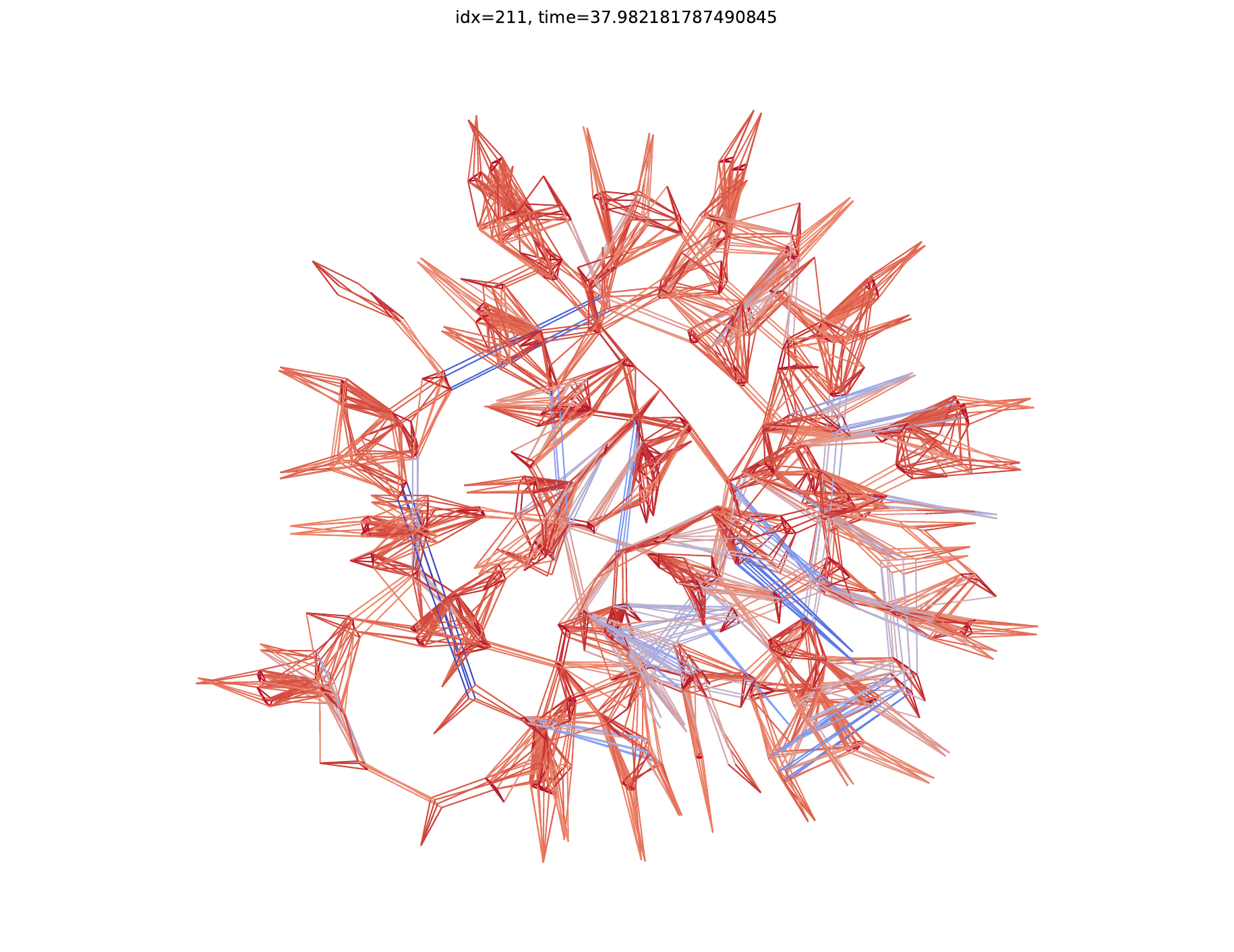} &
\imgcell{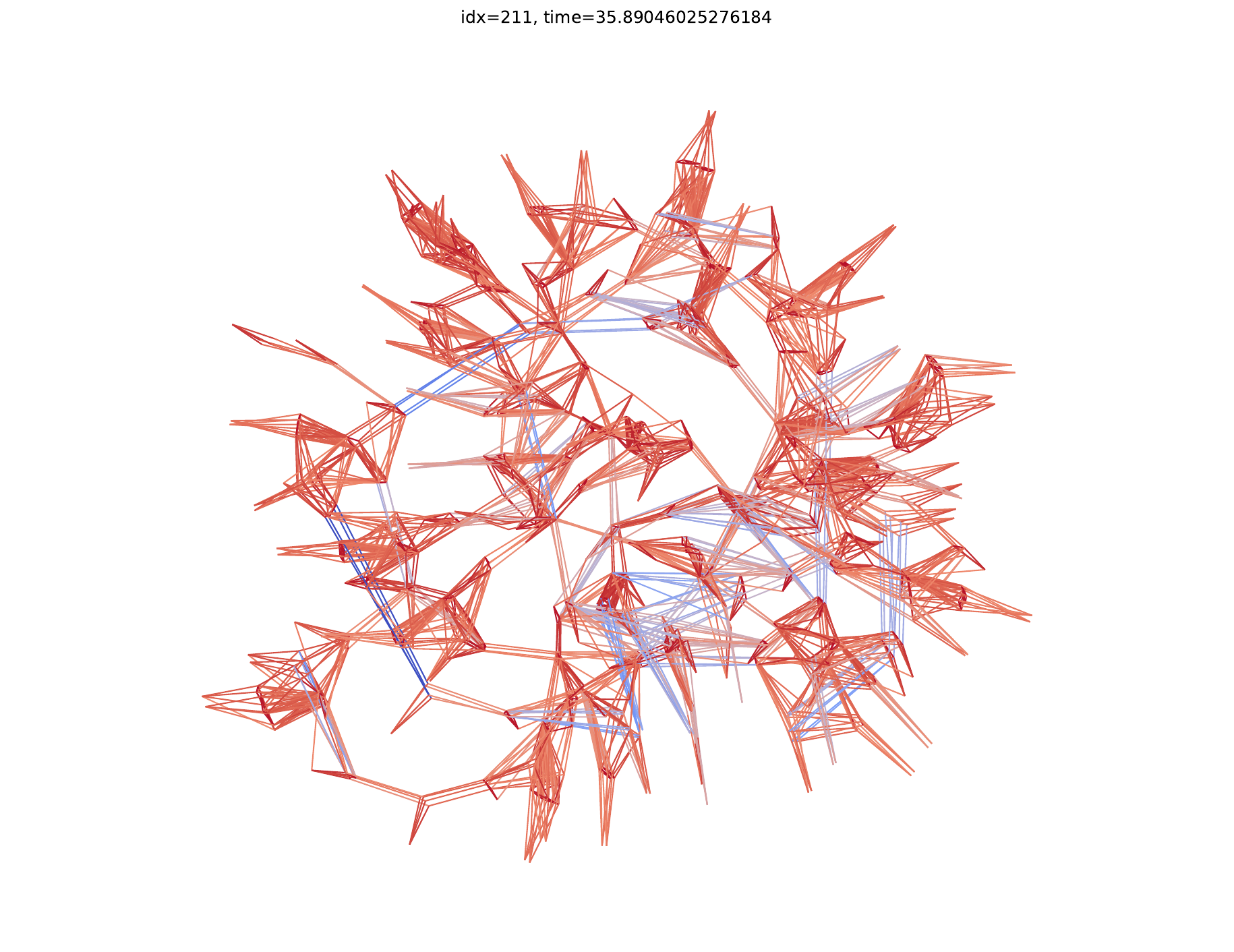} &
\imgcell{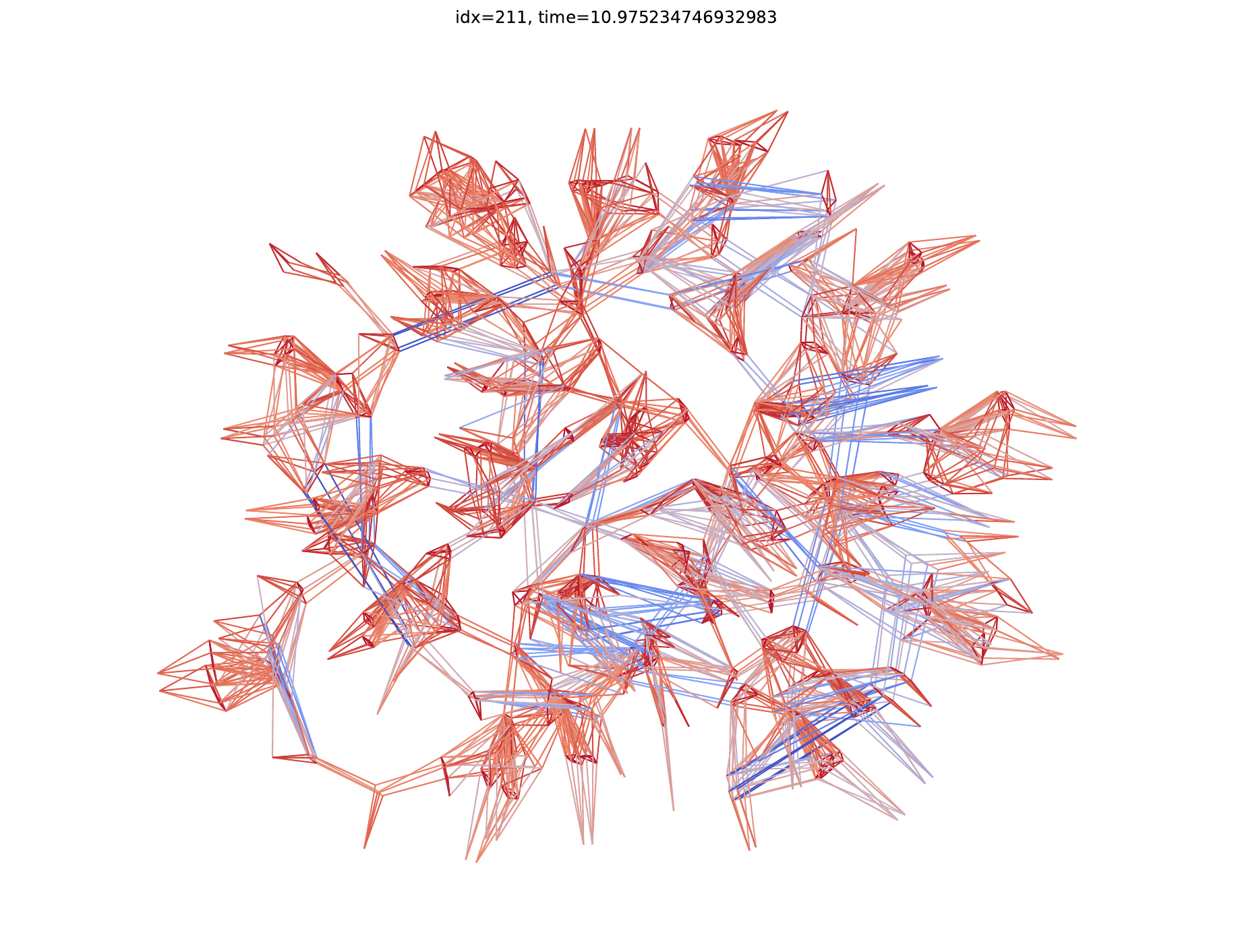} \\

&
t = 0.39s &
t = 18.72s &
t = 92.60s &
t = 3.88s &
t = 7200.00s &
t = 4.09s &
t = 3.13s &
t = 3.62s &
t = 3.19s &
t = 3.78s &
t = 3.59s &
t = 3.50s \\

\makecell{\bfseries radfr1\\N = 1048\\M = 12856} &
\imgcell{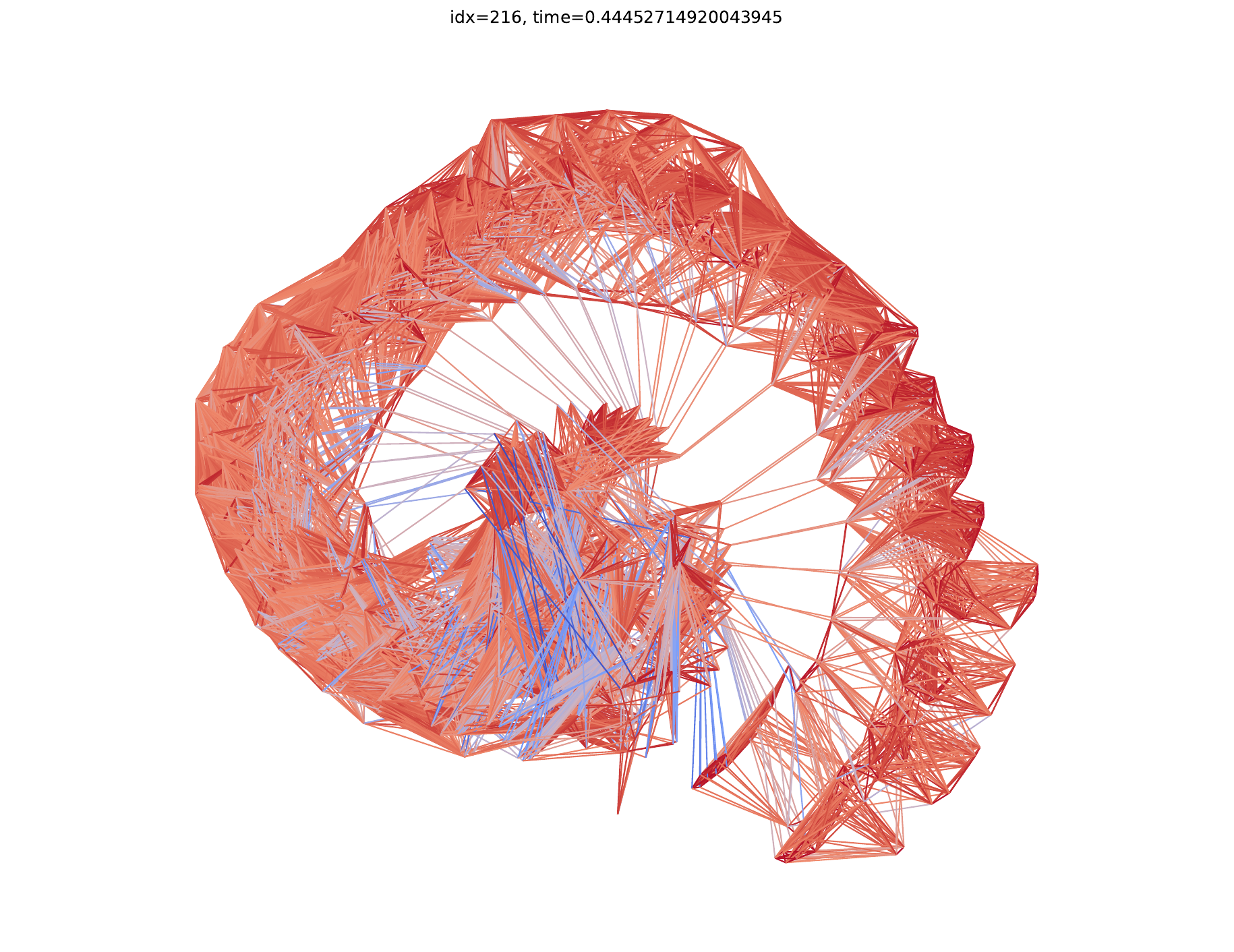} &
\imgcell{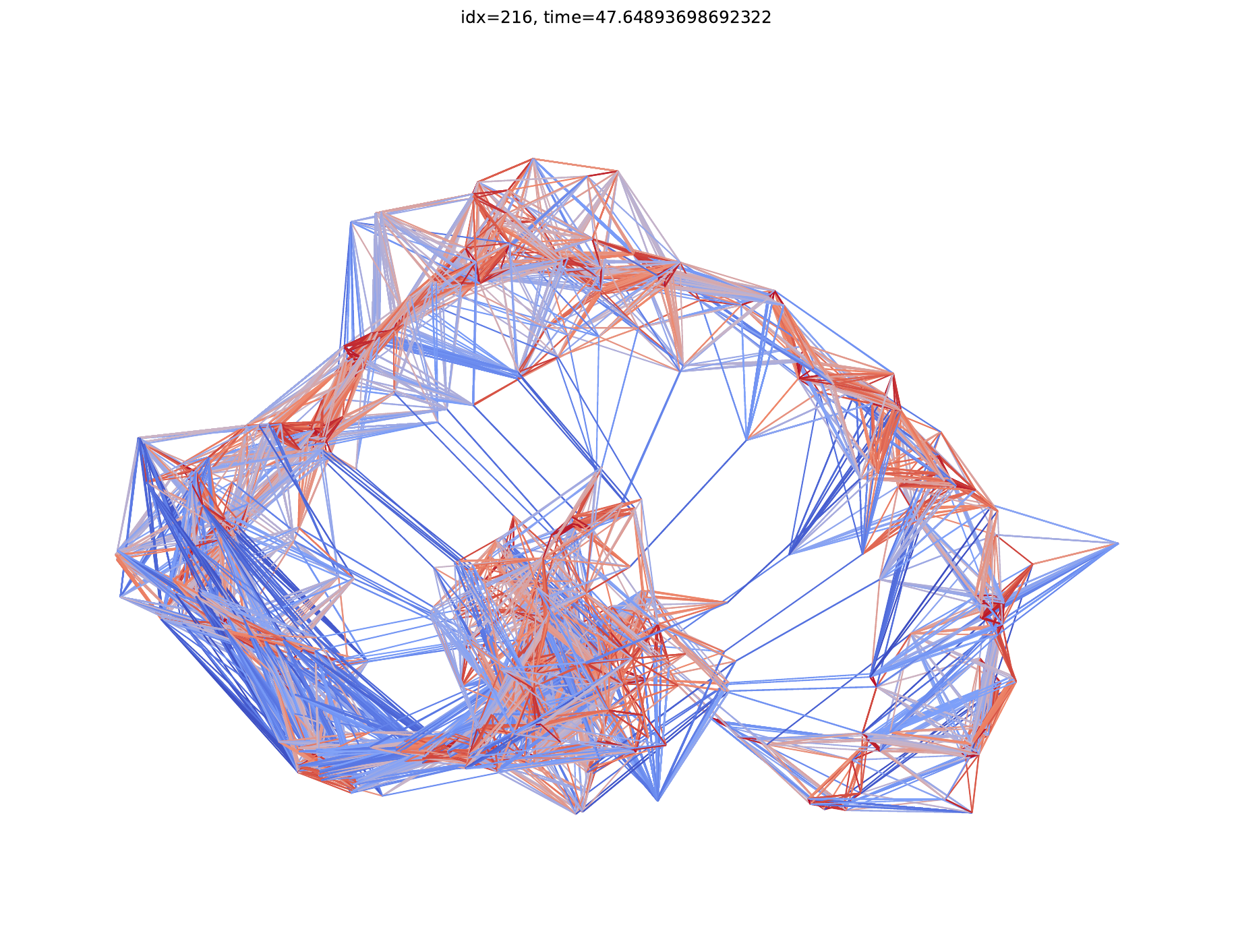} &
\imgcell{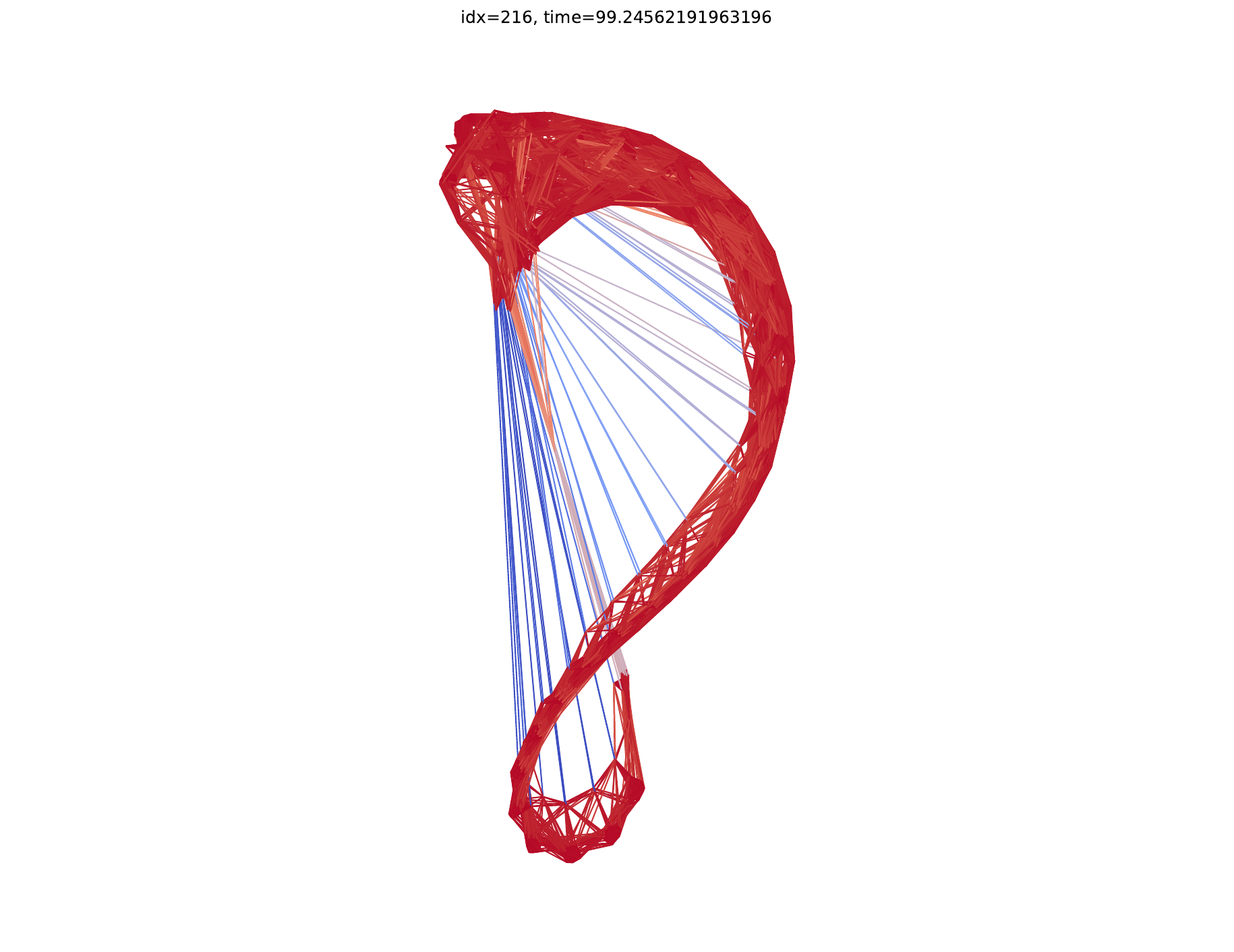} &
\imgcell{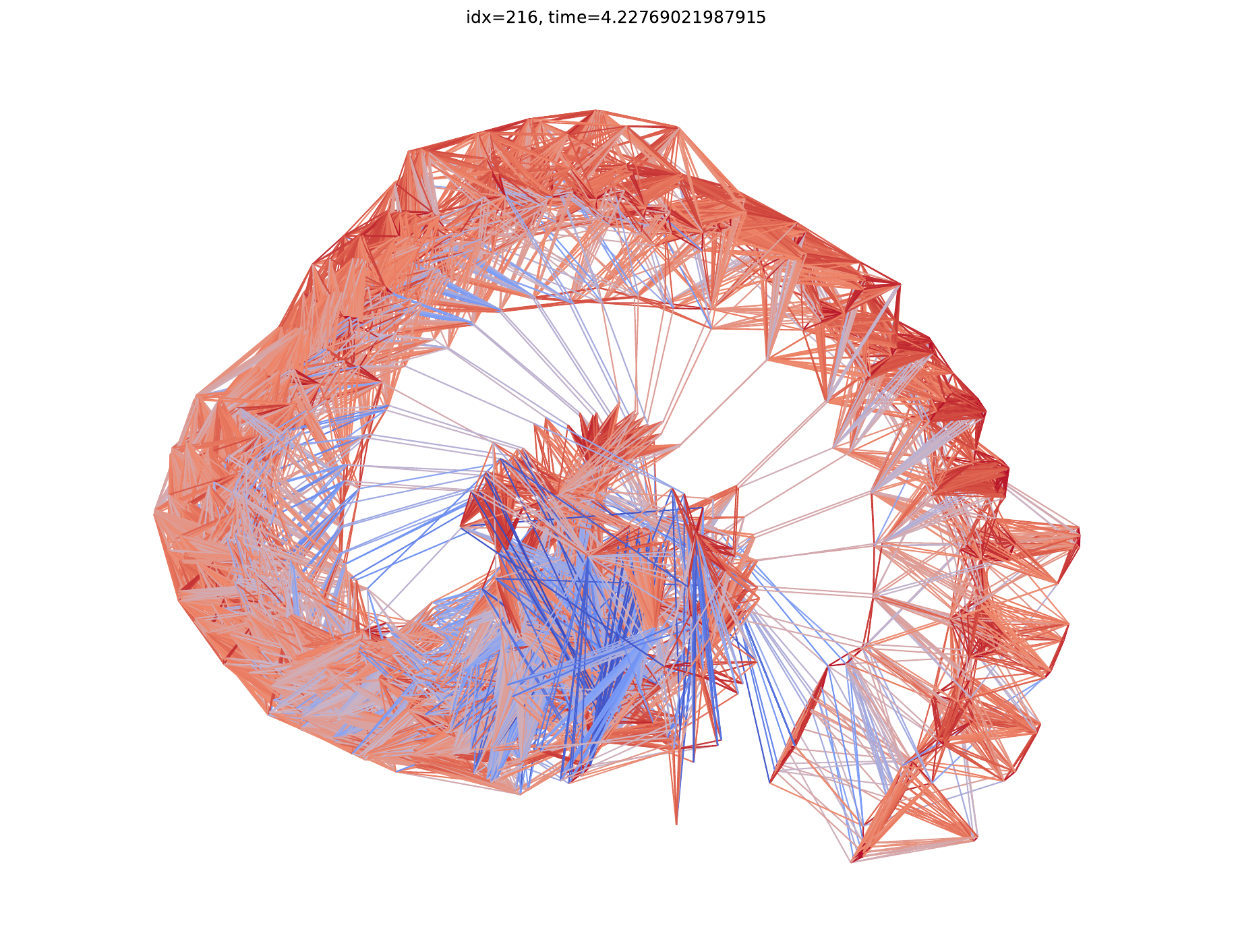} &
\imgcell{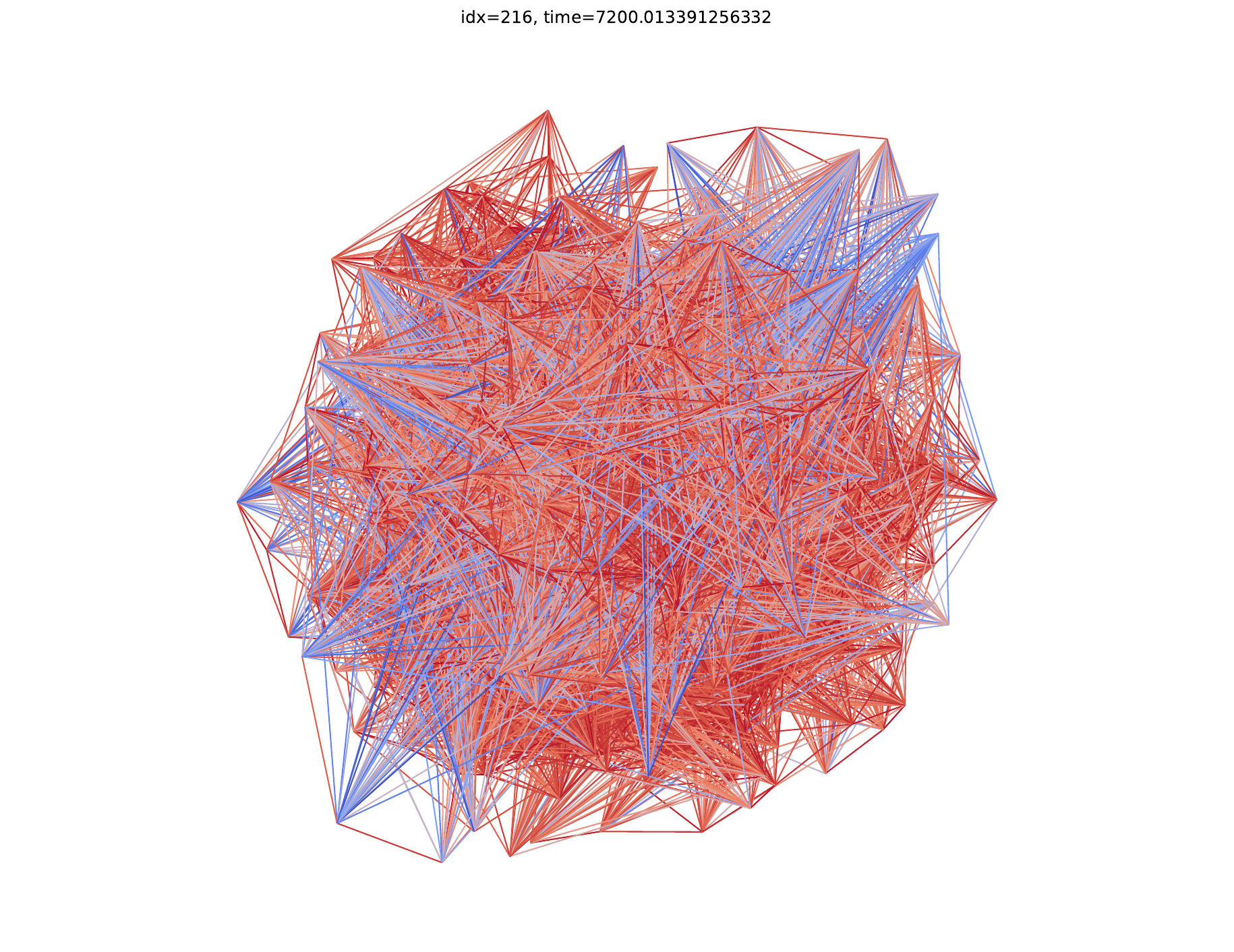} &
\imgcell{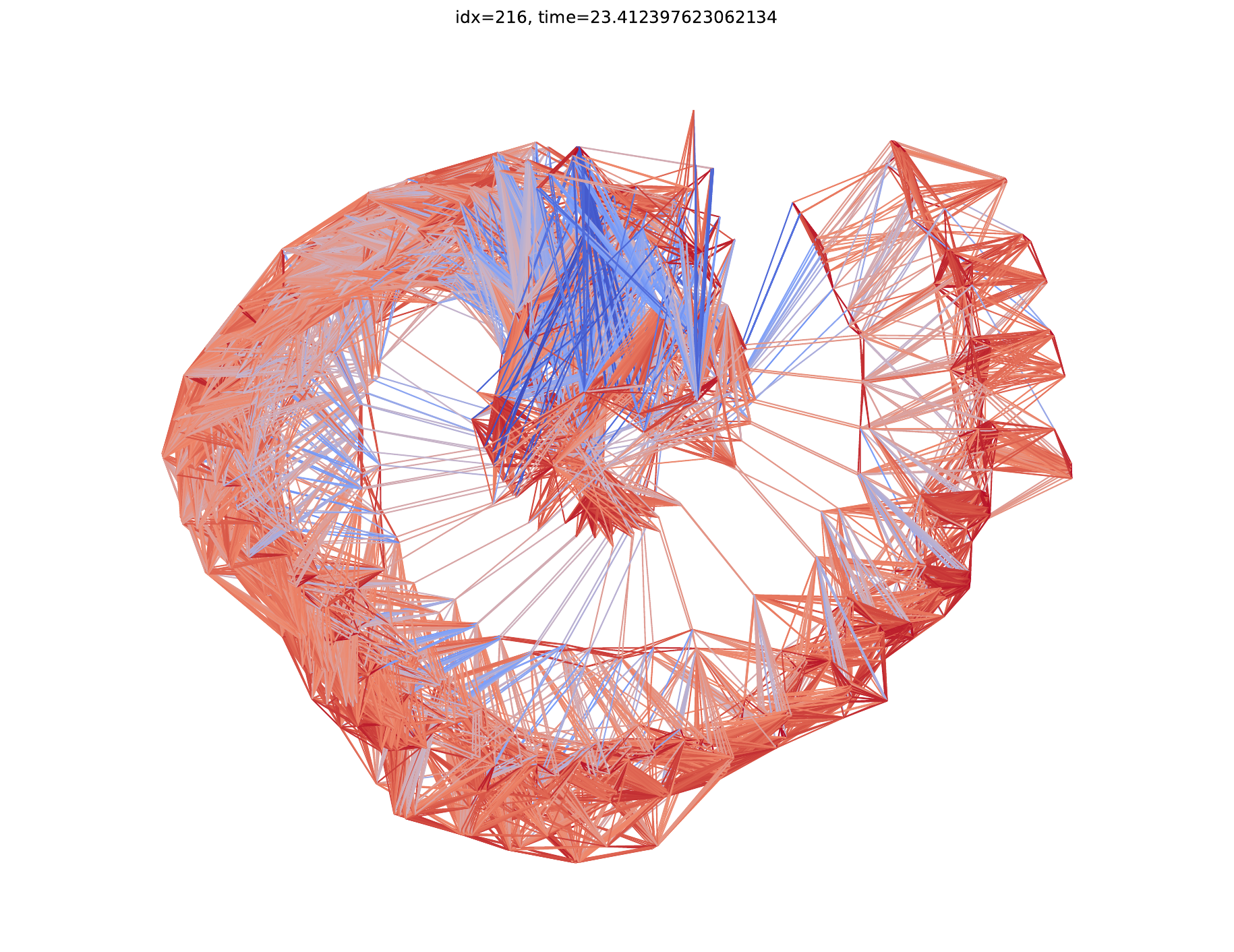} &
\imgcell{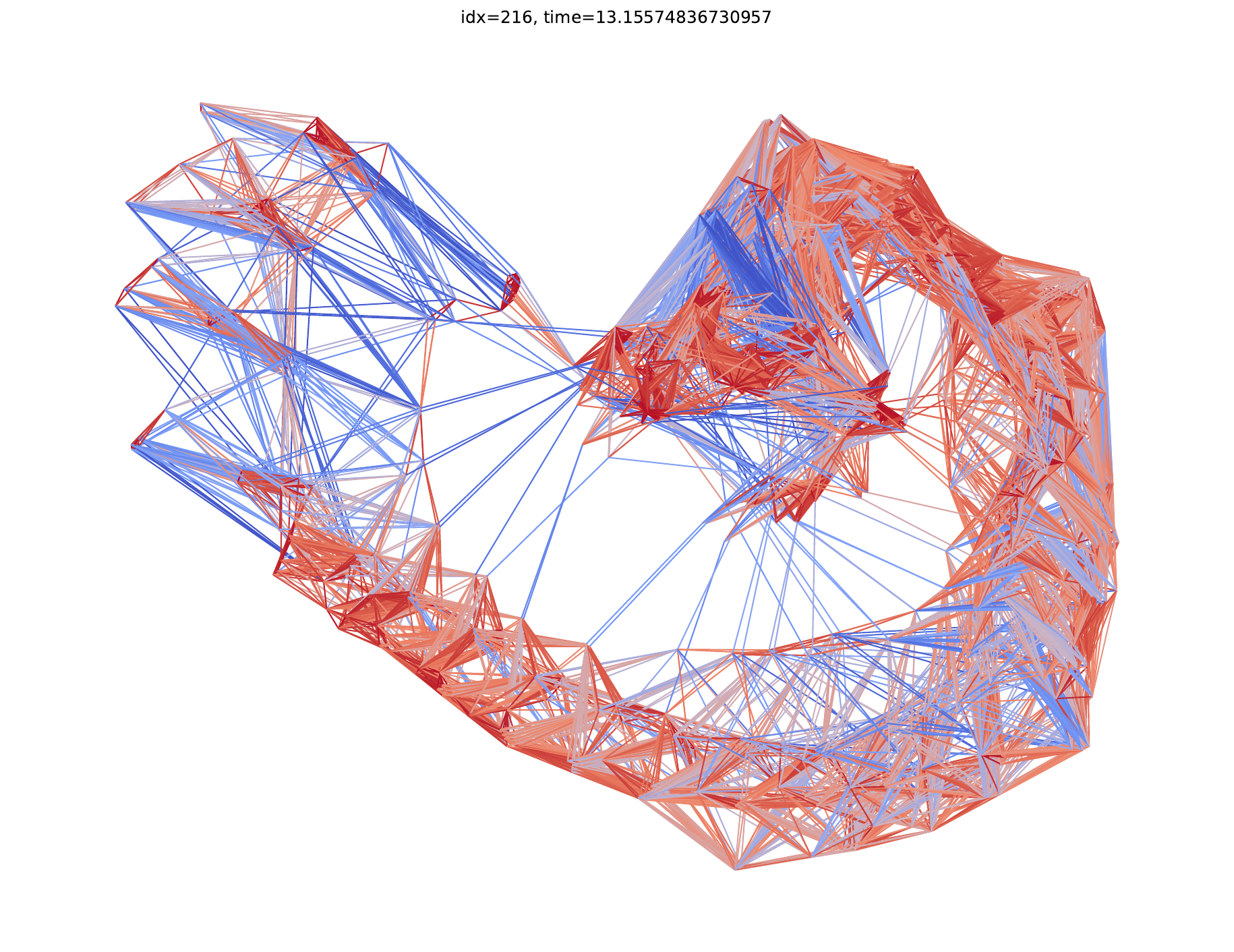} &
\imgcell{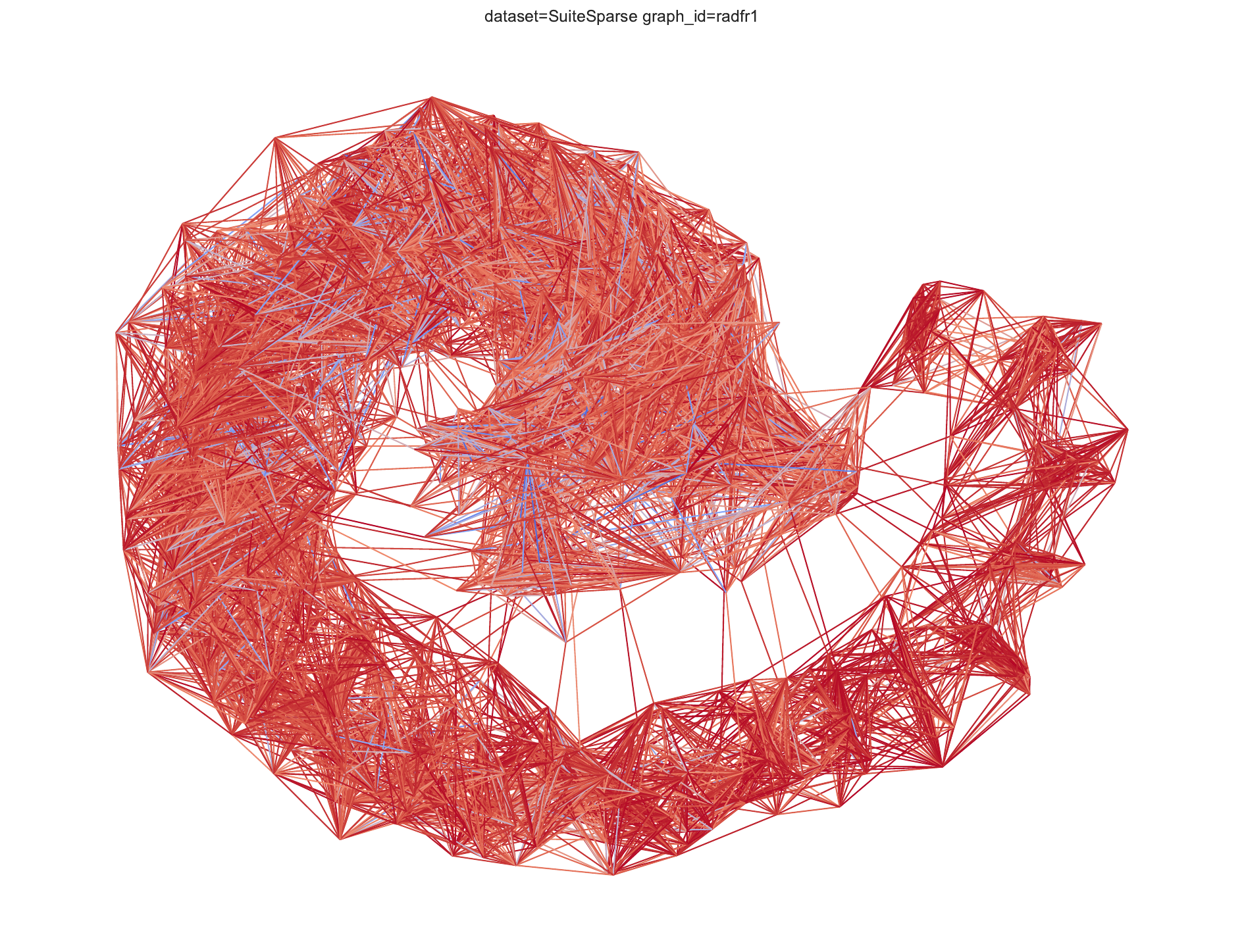} &
\imgcell{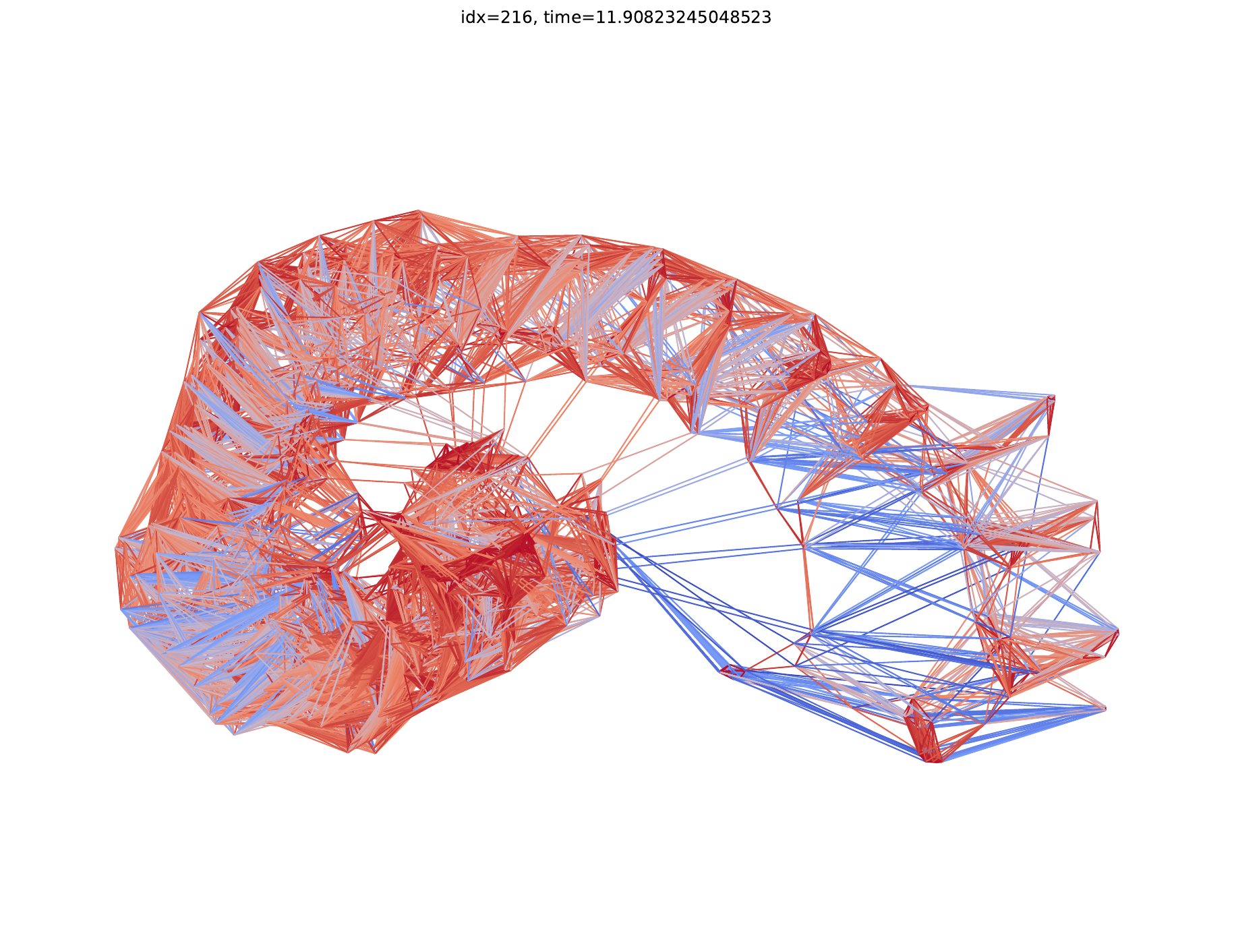} &
\imgcell{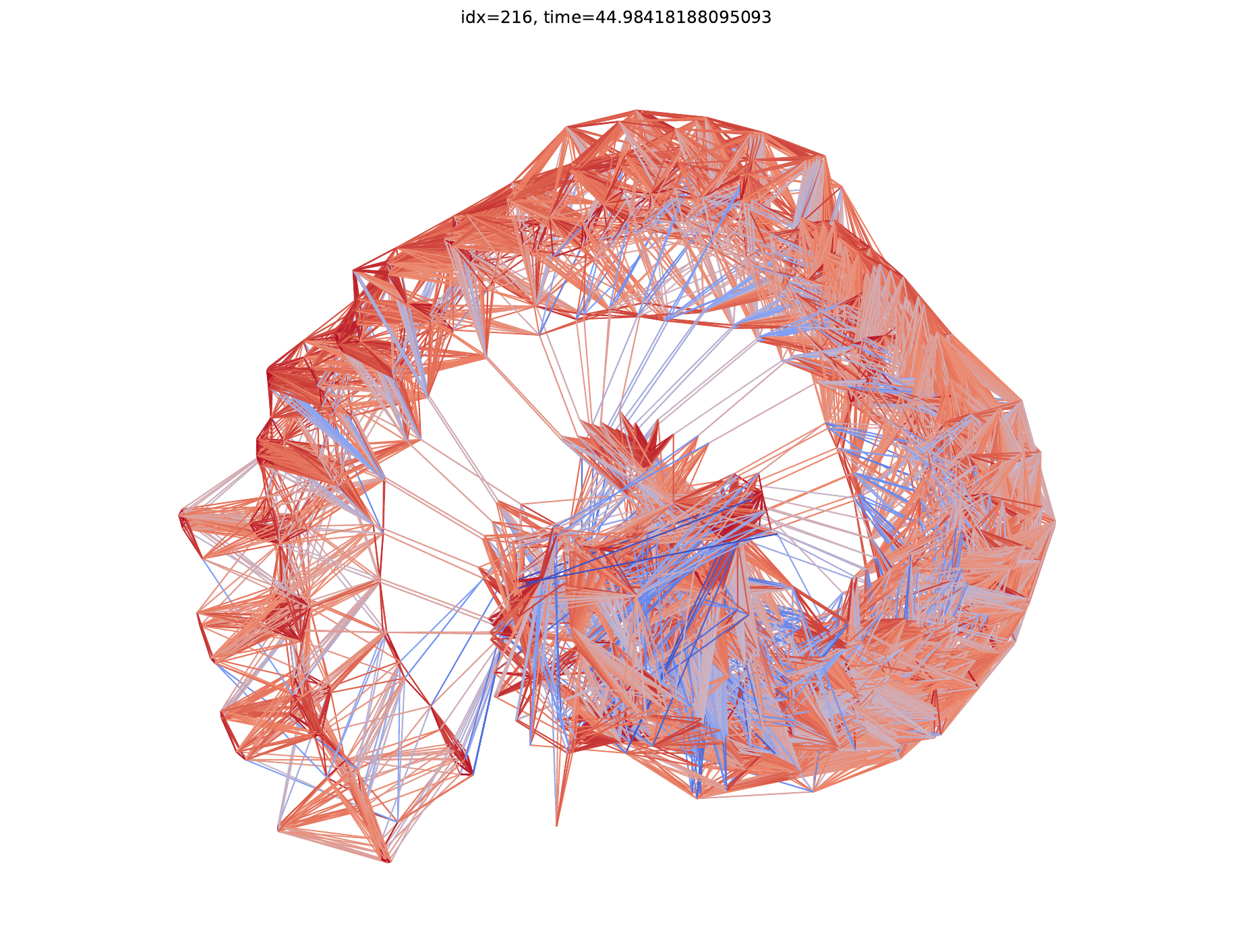} &
\imgcell{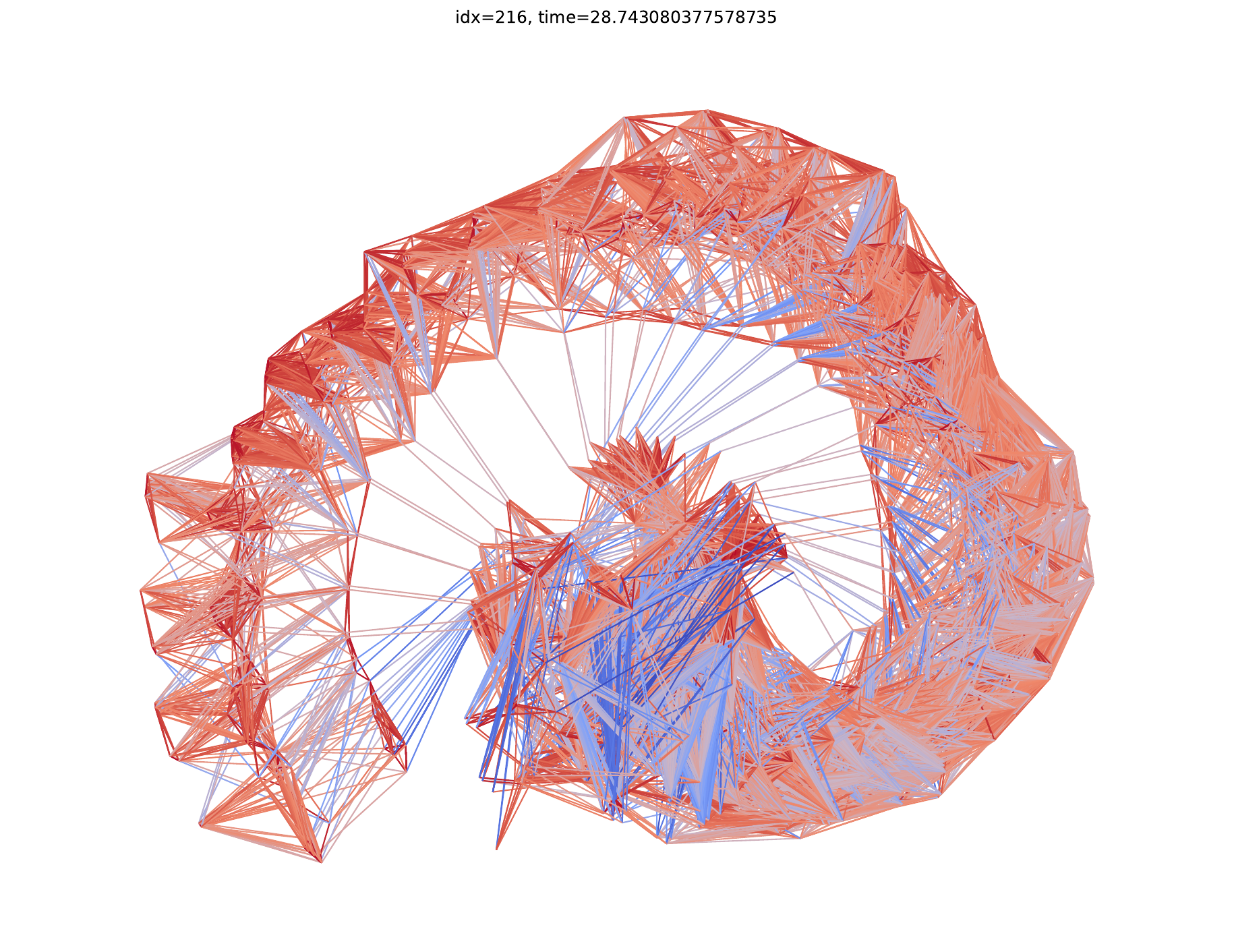} &
\imgcell{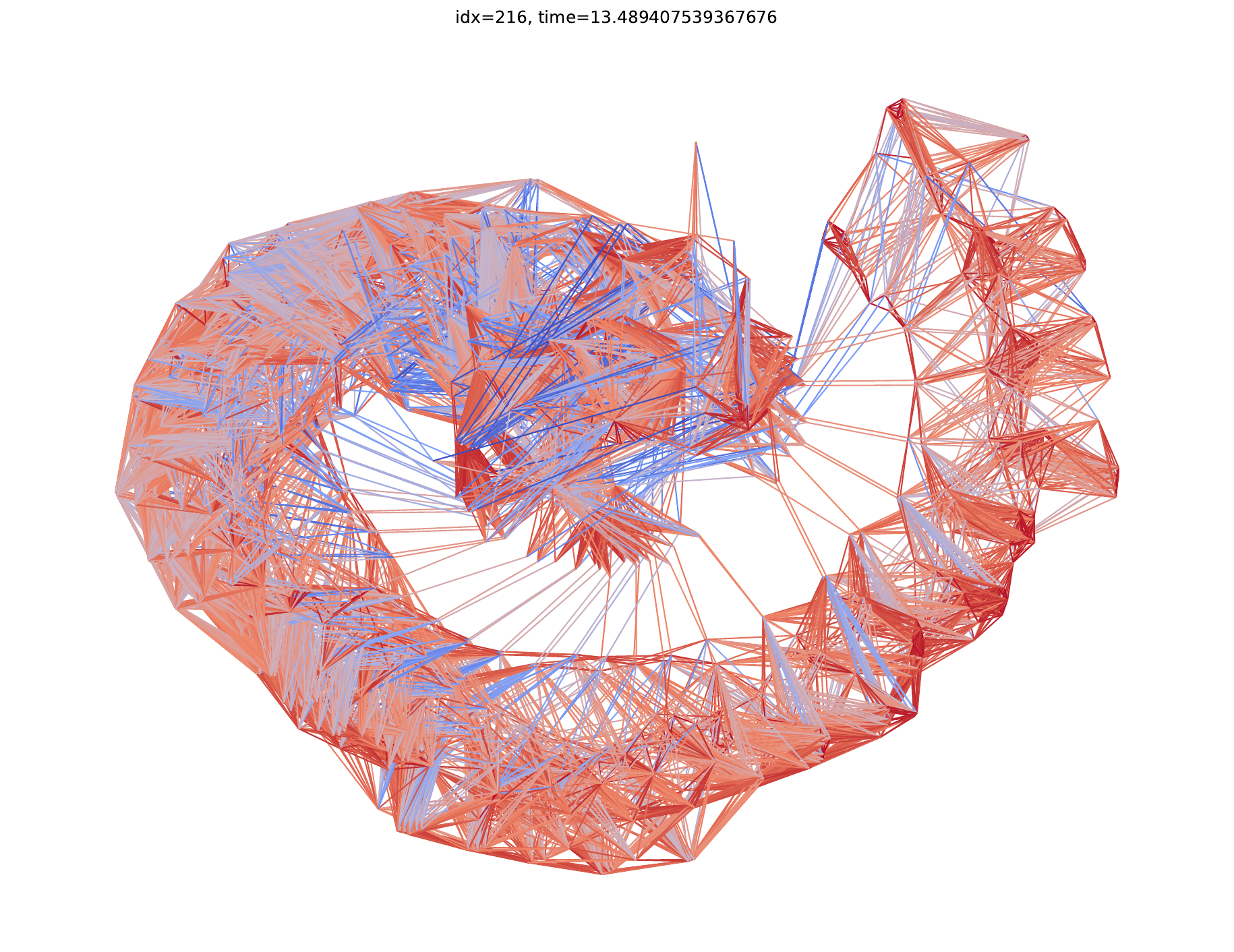} \\

&
t = 0.44s &
t = 47.65s &
t = 99.25s &
t = 4.23s &
t = 7200.00s &
t = 4.14s &
t = 4.27s &
t = 4.54s &
t = 4.12s &
t = 4.48s &
t = 3.84s &
t = 4.39s \\

\makecell{\bfseries can\_1054\\N = 1054\\M = 5571} &
\imgcell{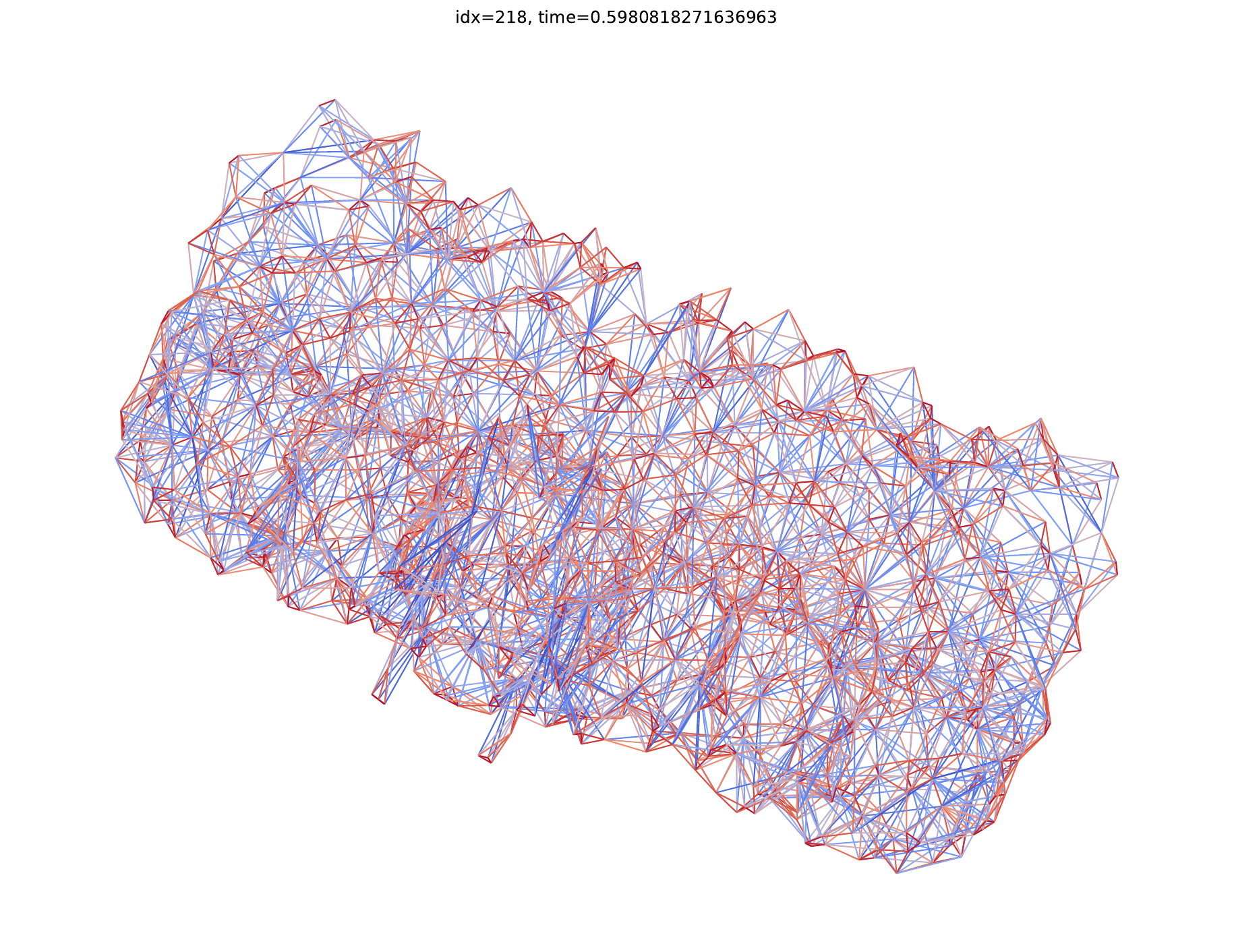} &
\imgcell{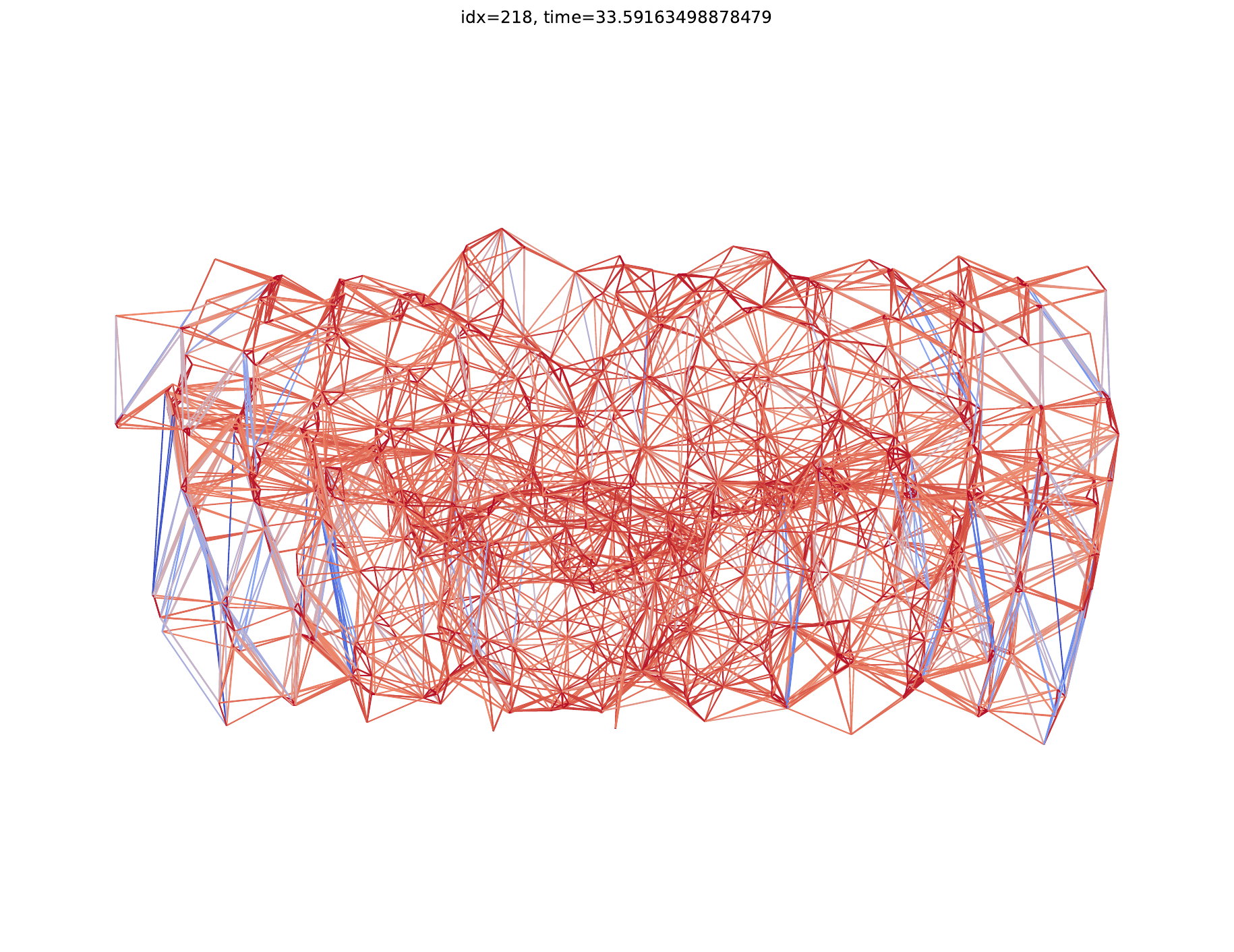} &
\imgcell{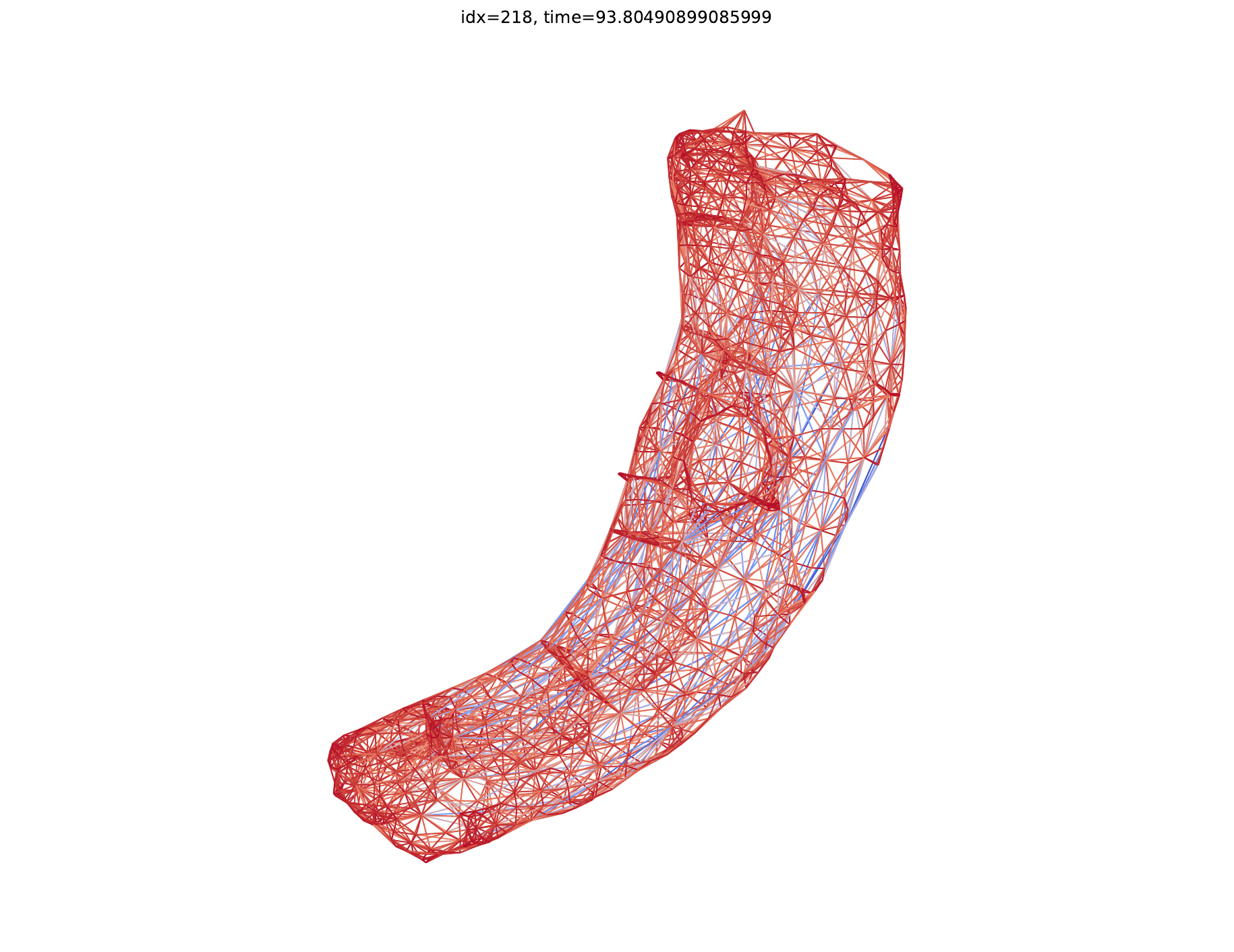} &
\imgcell{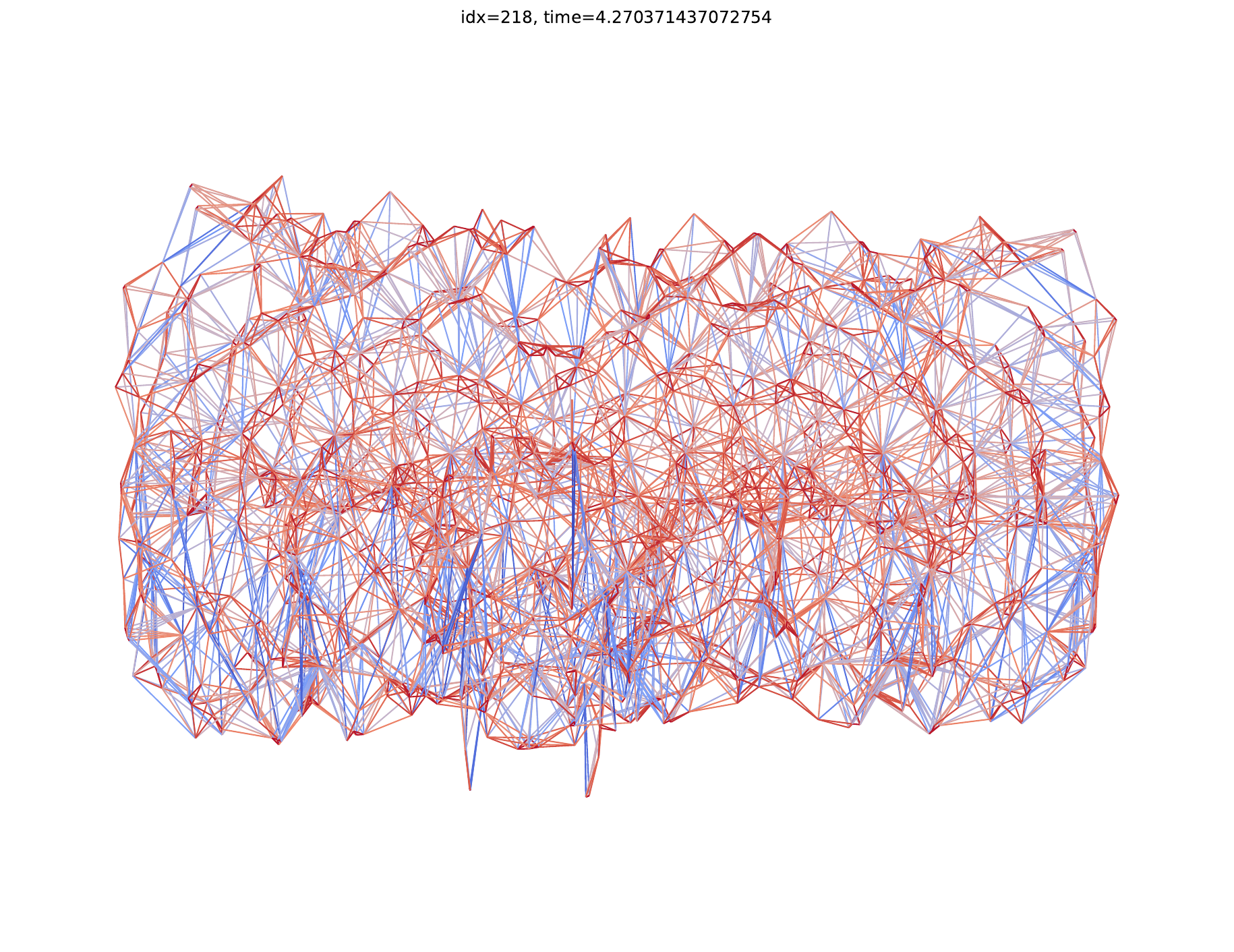} &
\imgcell{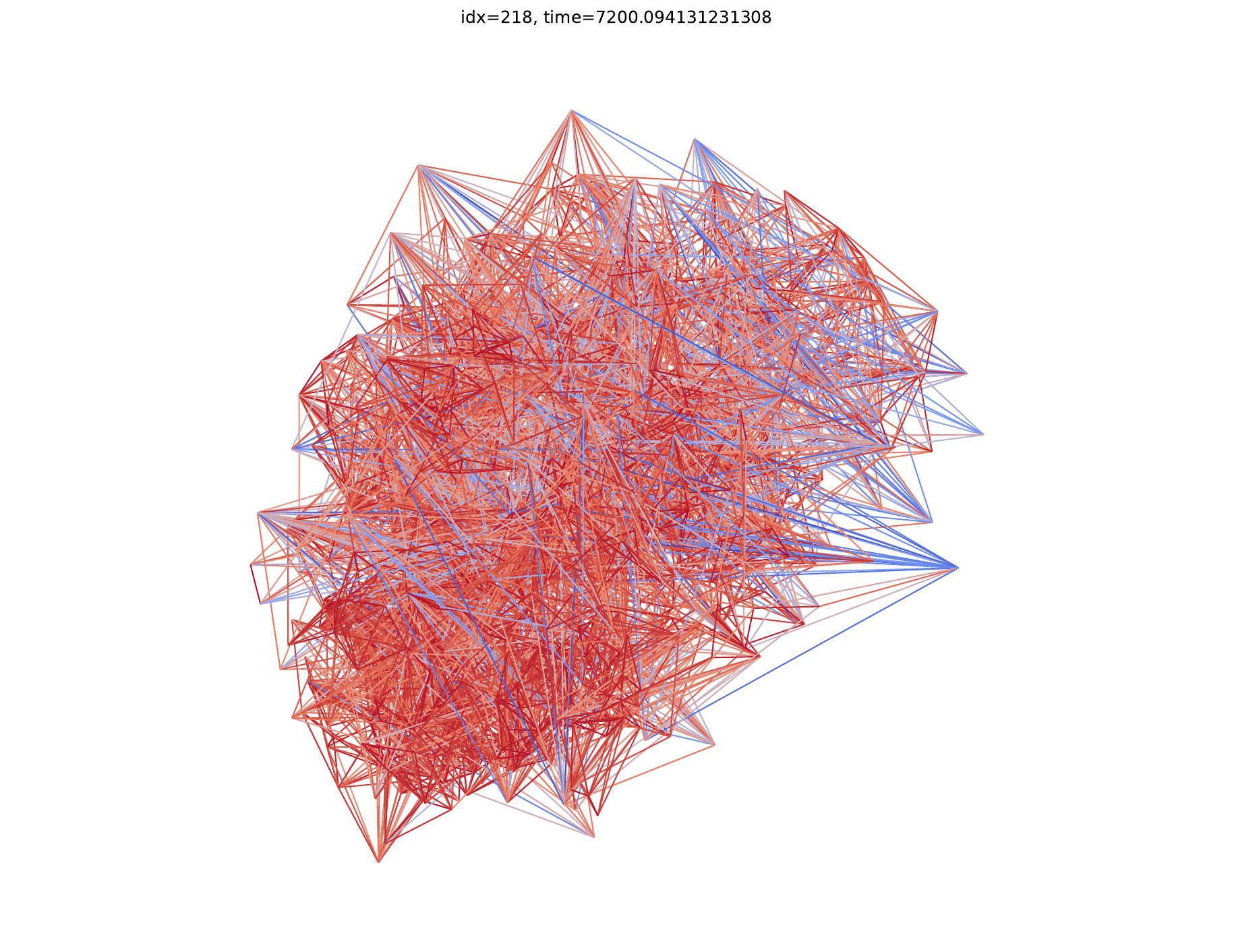} &
\imgcell{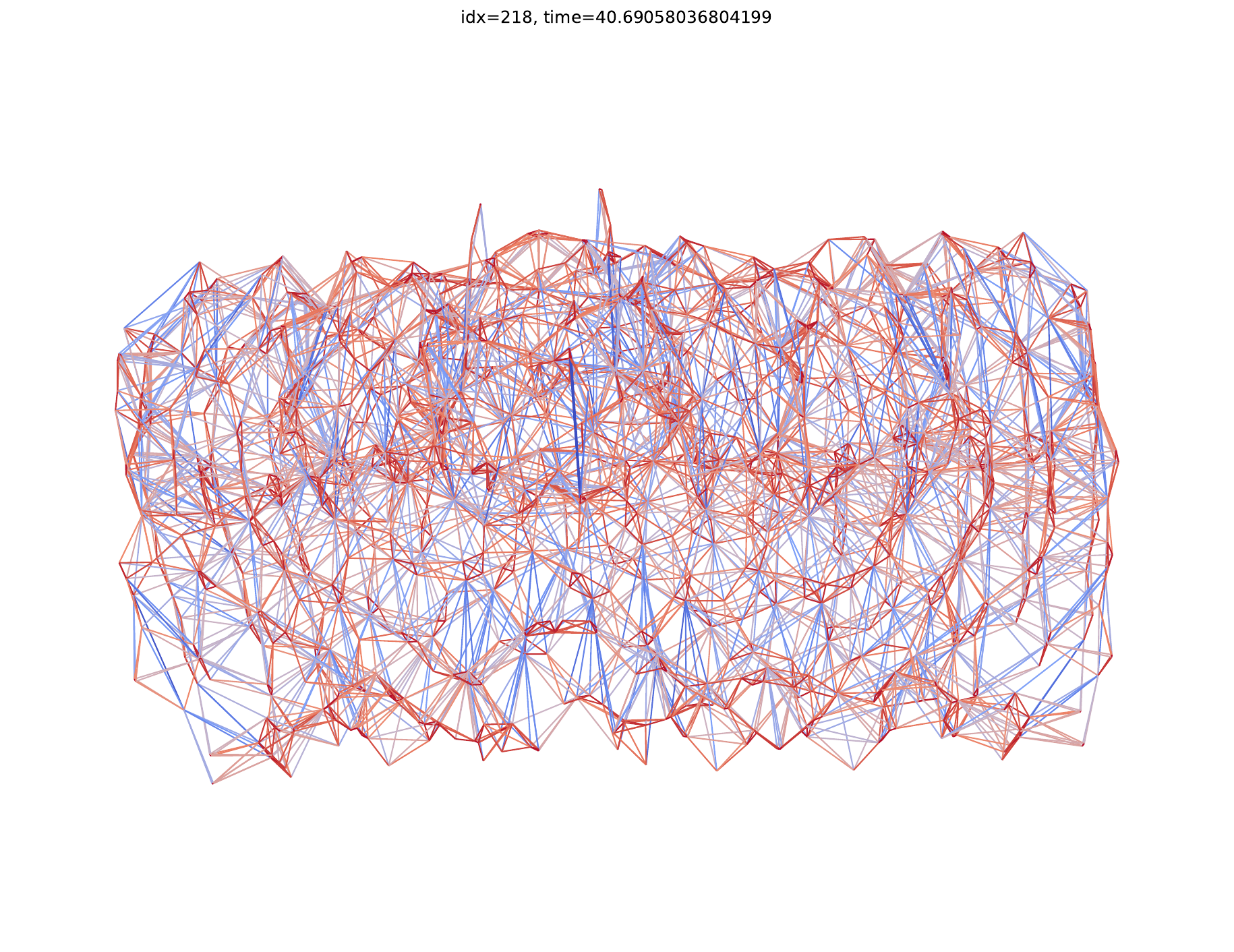} &
\imgcell{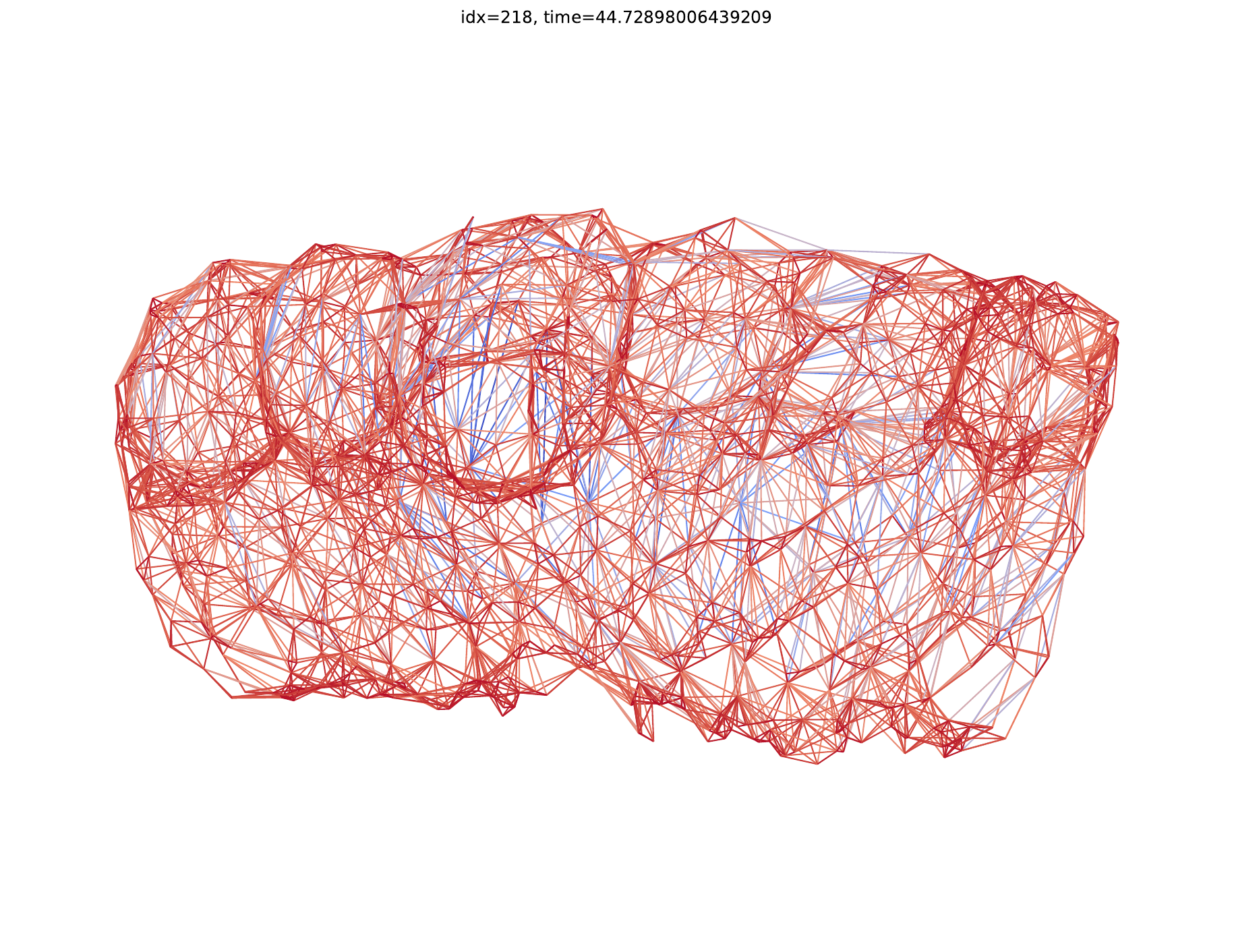} &
\imgcell{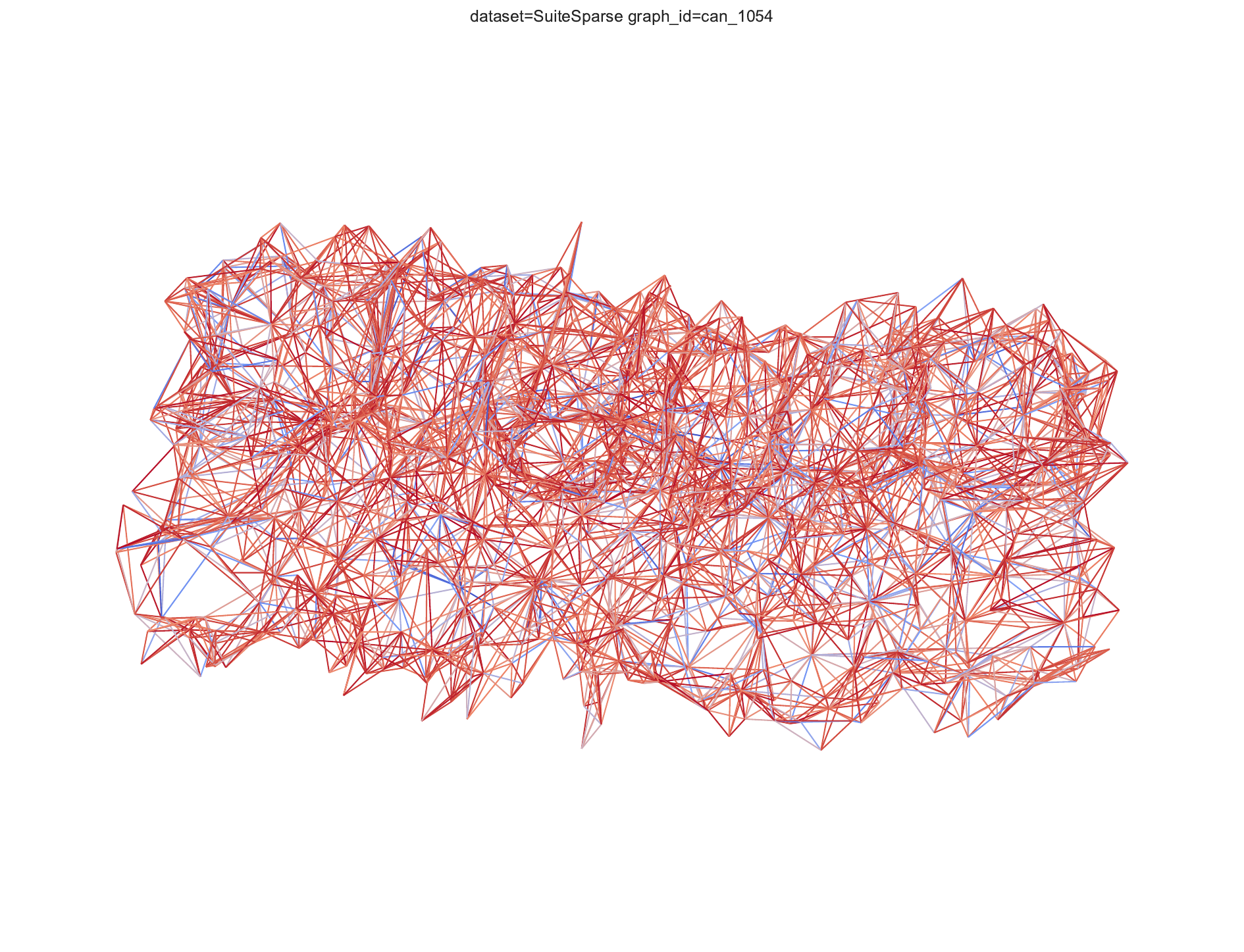} &
\imgcell{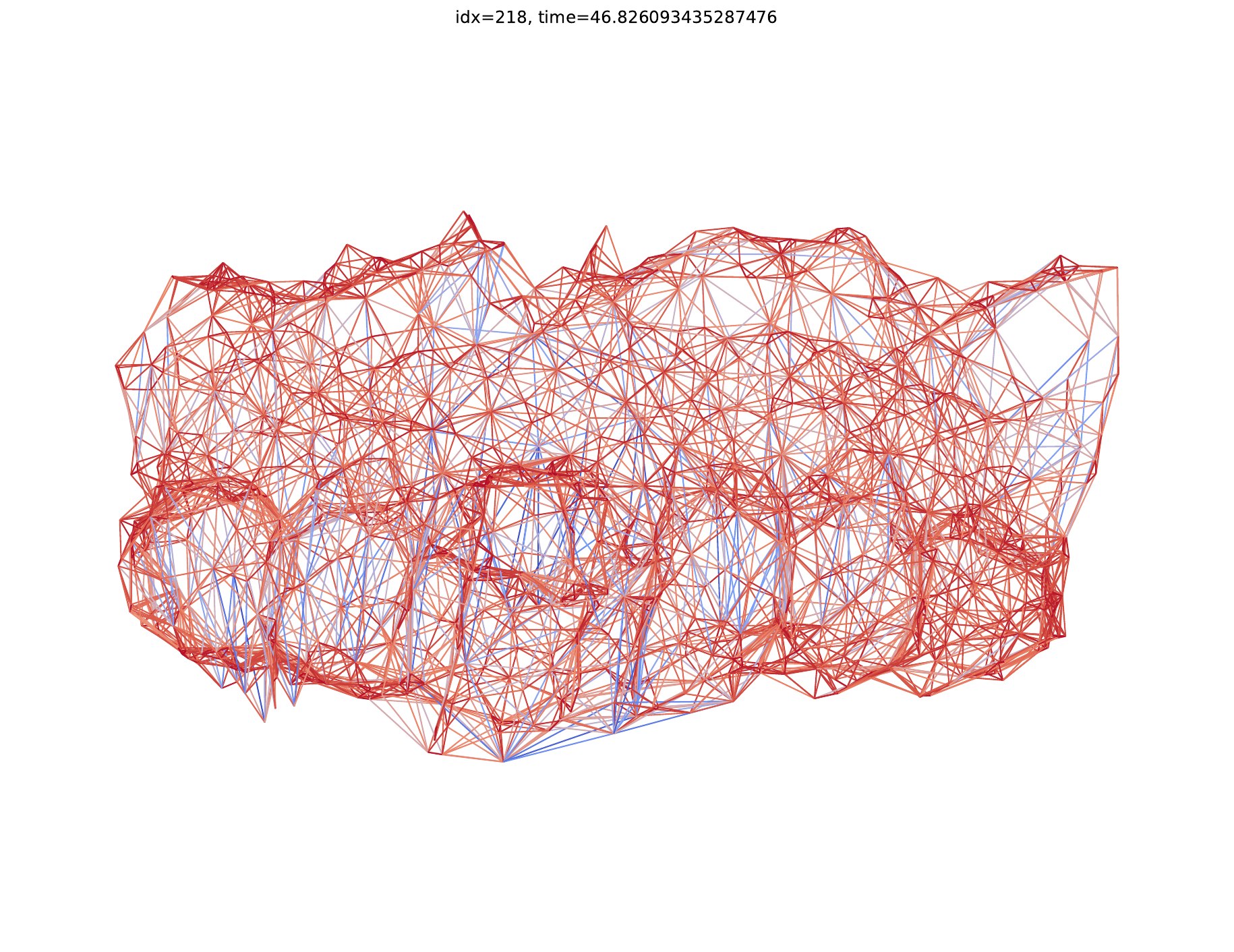} &
\imgcell{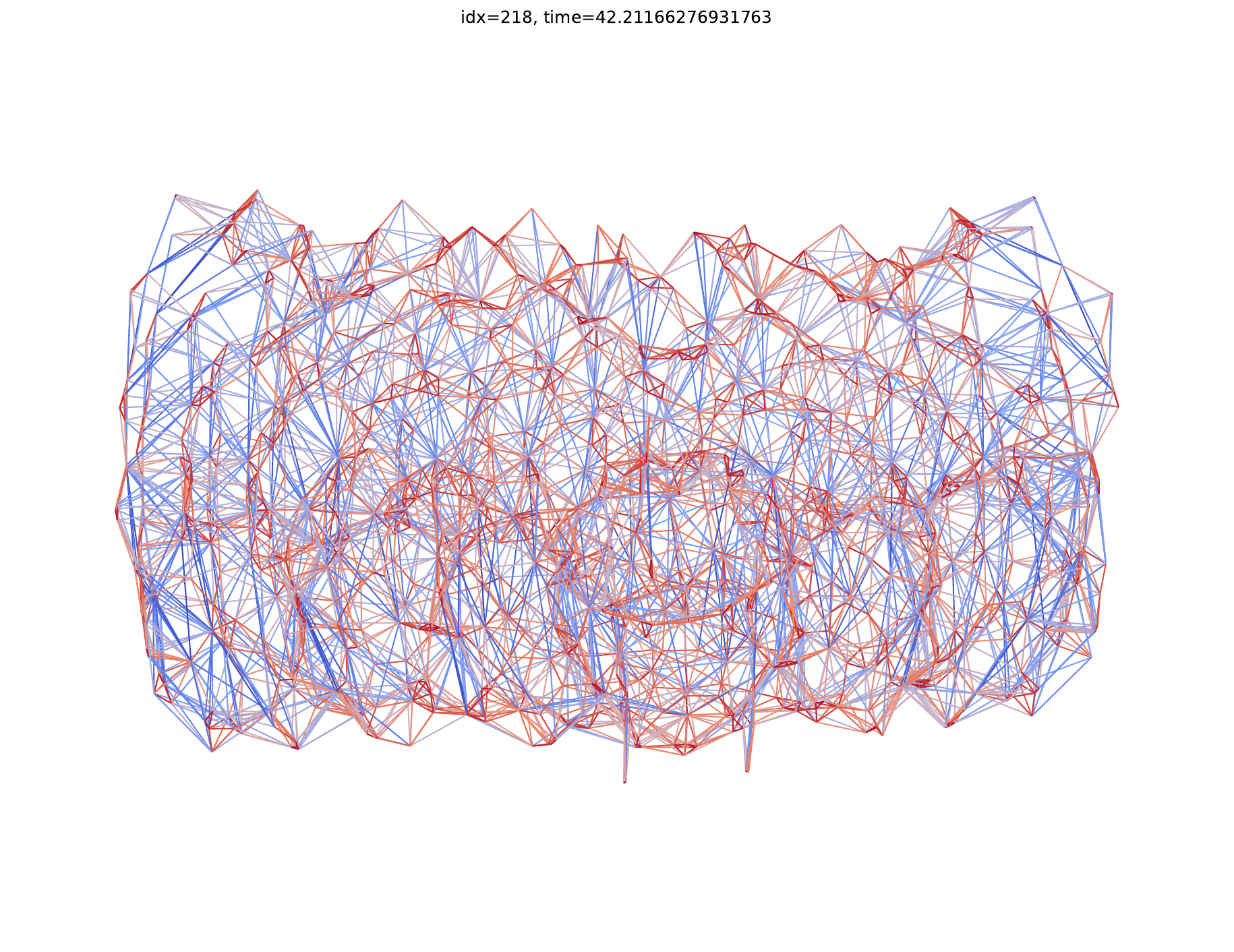} &
\imgcell{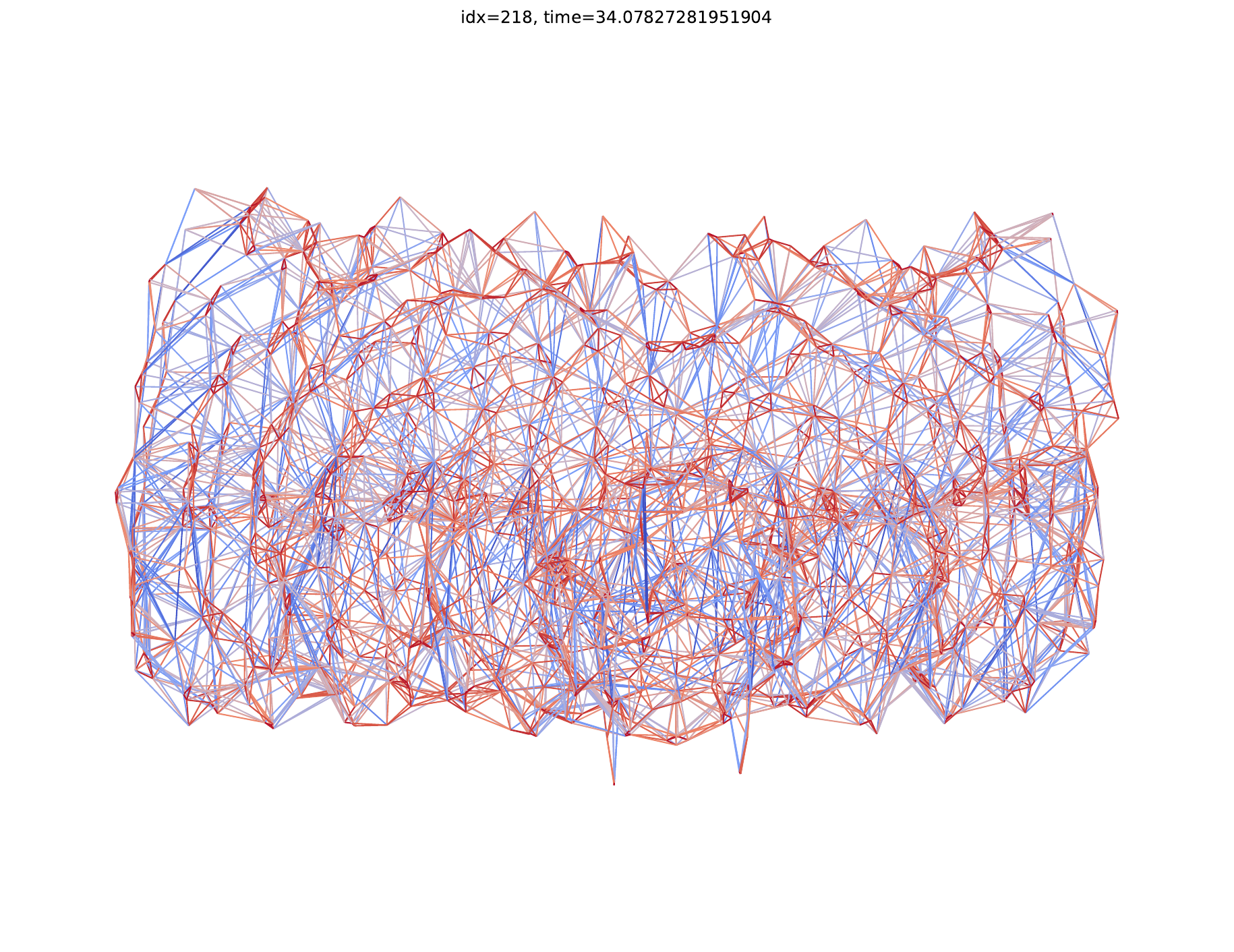} &
\imgcell{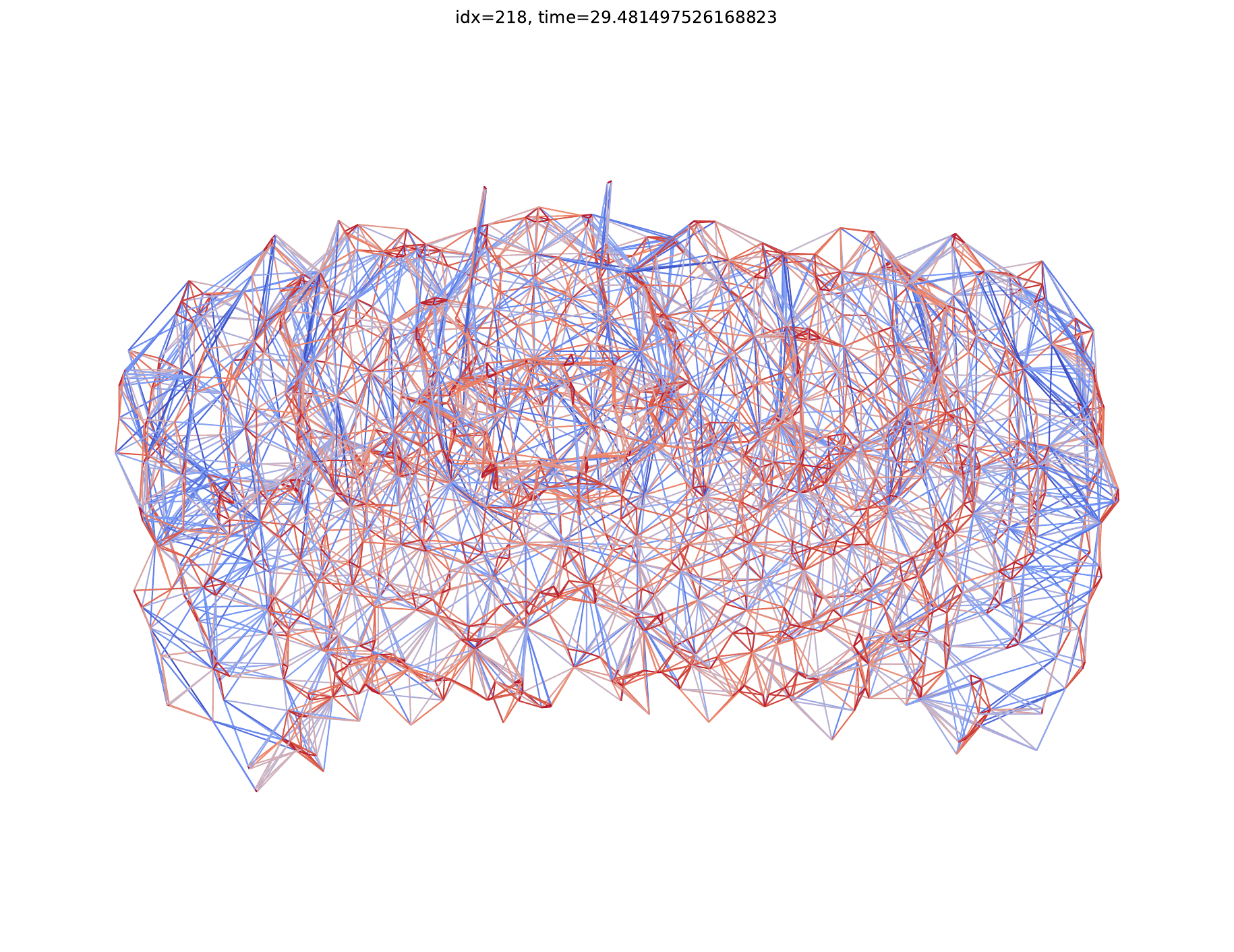} \\

&
t = 0.60s &
t = 33.59s &
t = 93.80s &
t = 4.27s &
t = 7200.00s &
t = 4.09s &
t = 3.43s &
t = 4.15s &
t = 3.63s &
t = 4.21s &
t = 3.48s &
t = 2.98s \\

\makecell{\bfseries can\_1072\\N = 1072\\M = 5686} &
\imgcell{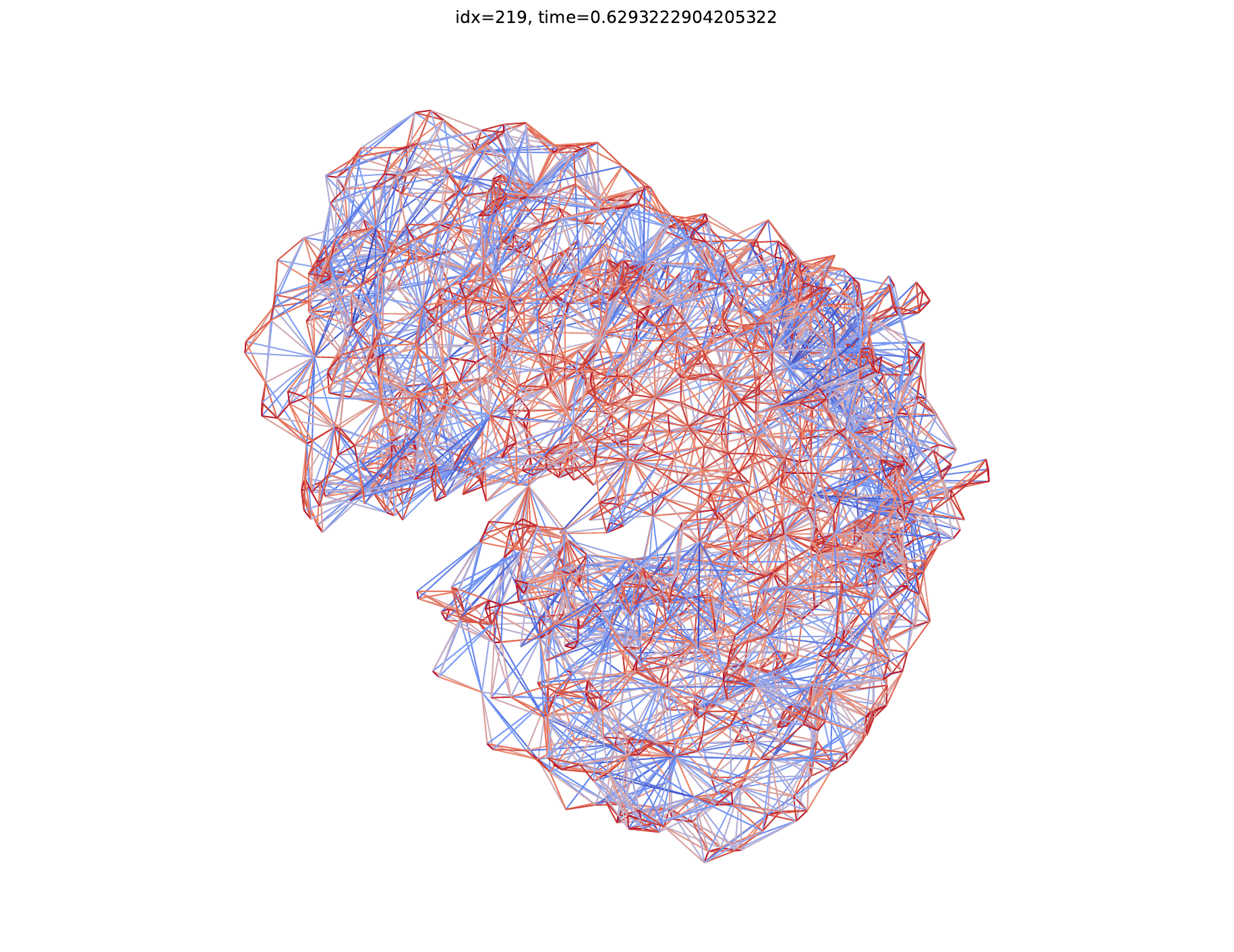} &
\imgcell{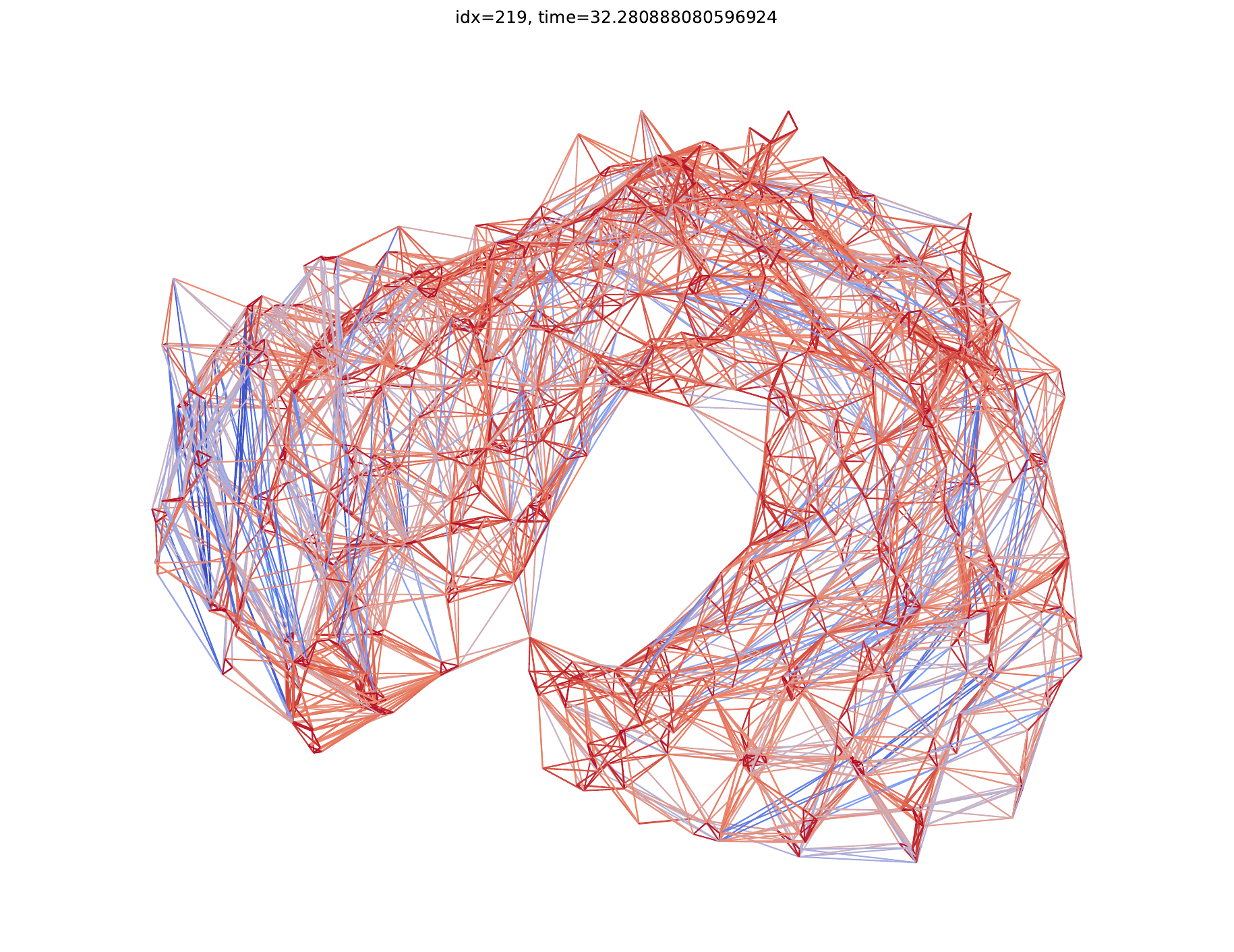} &
\imgcell{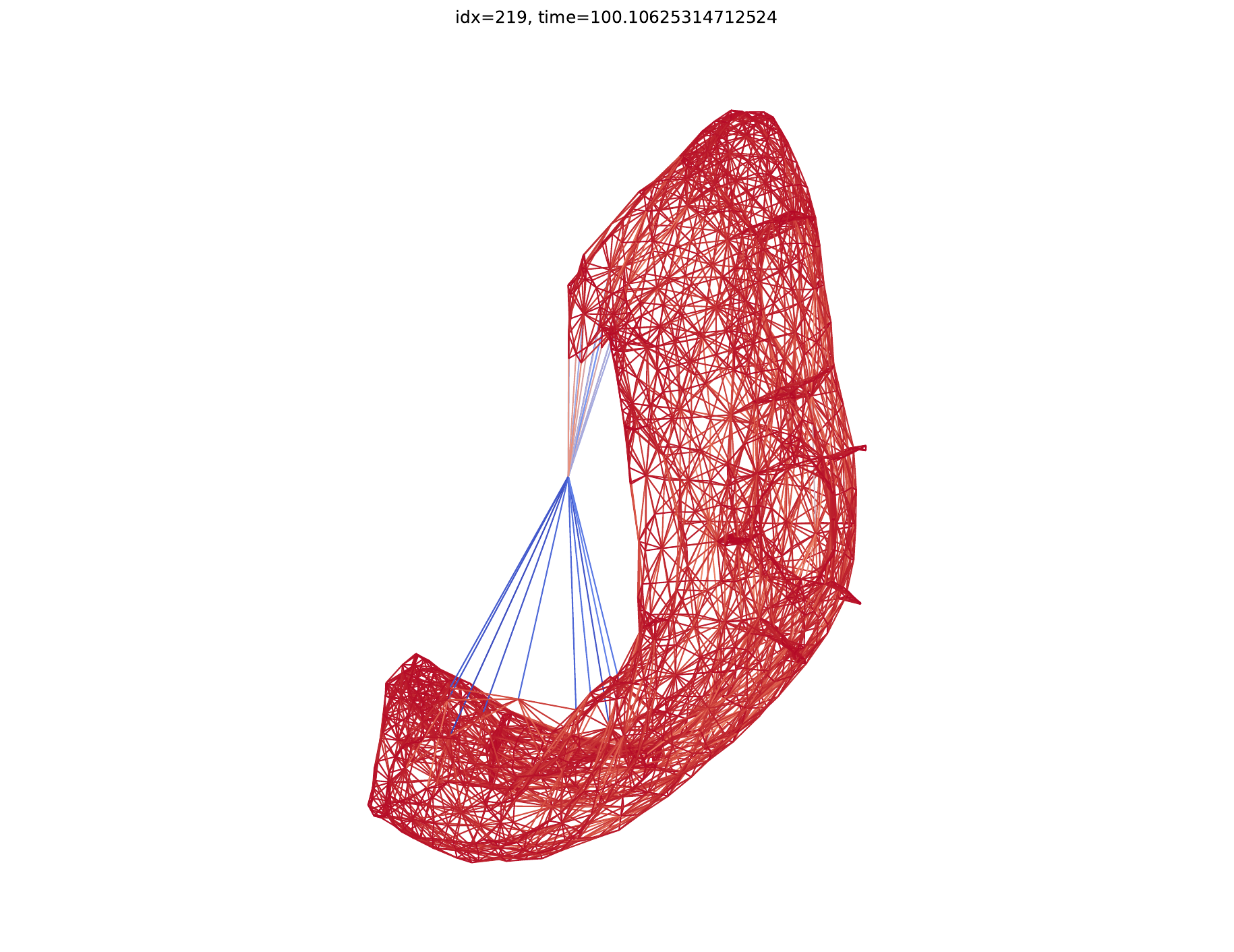} &
\imgcell{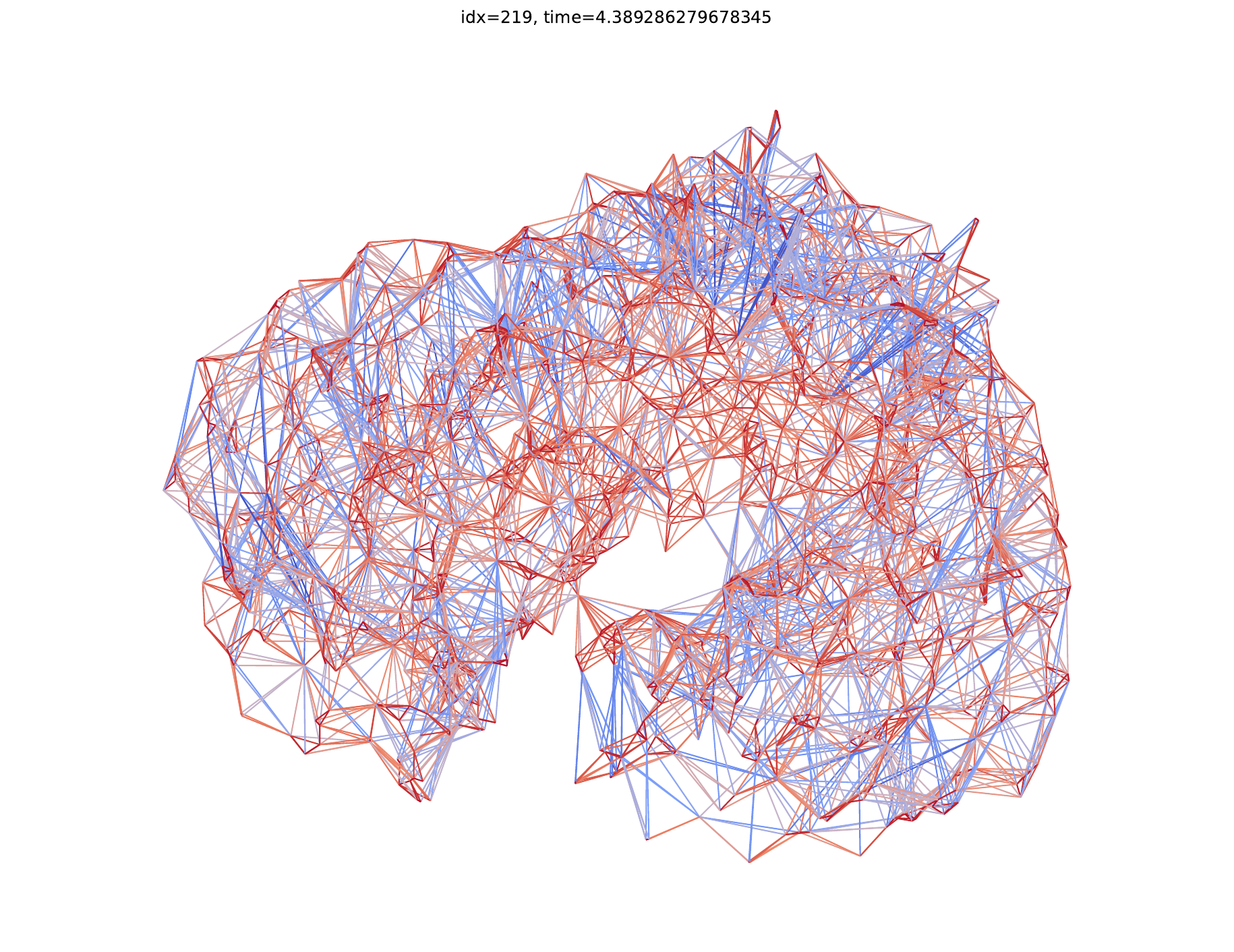} &
\imgcell{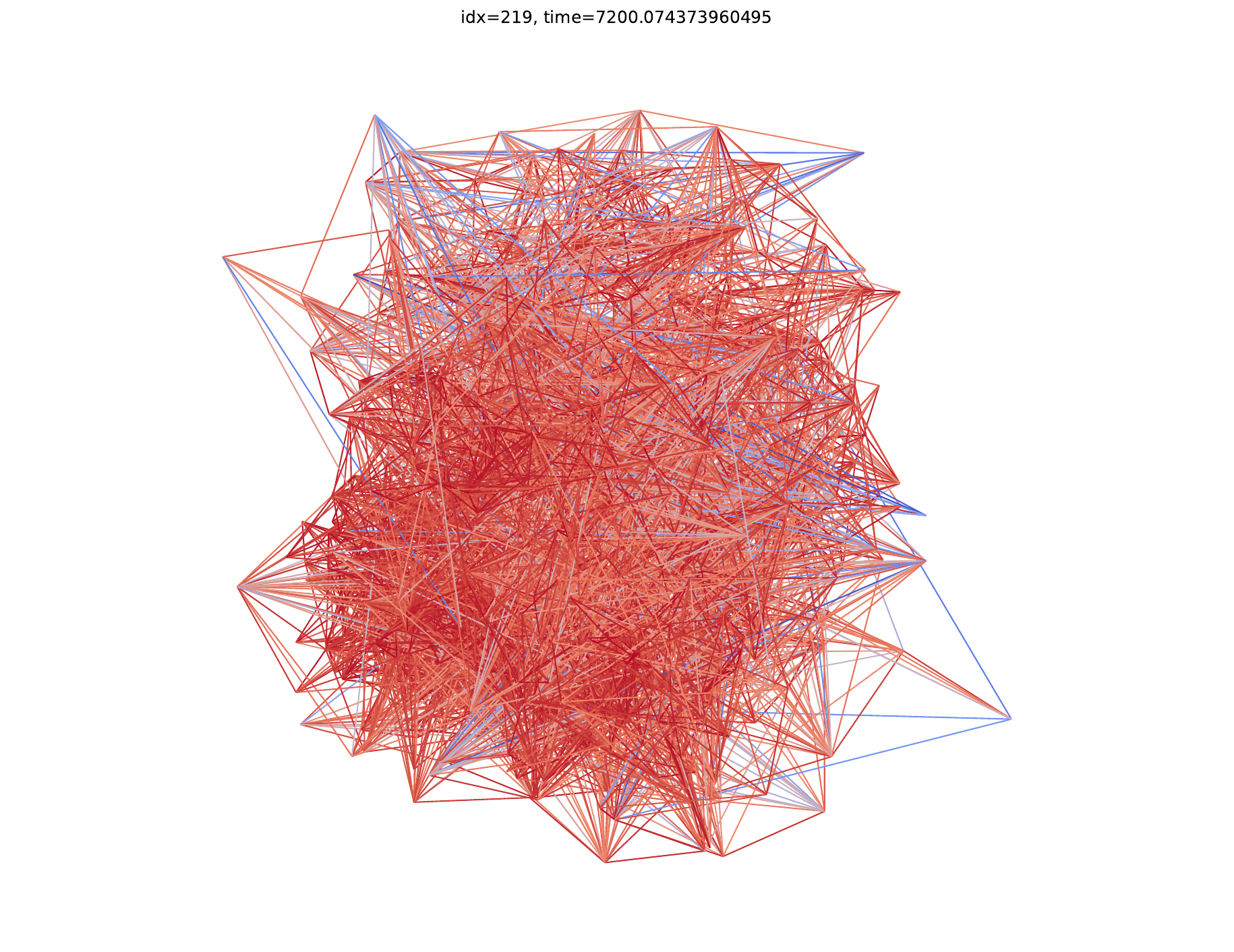} &
\imgcell{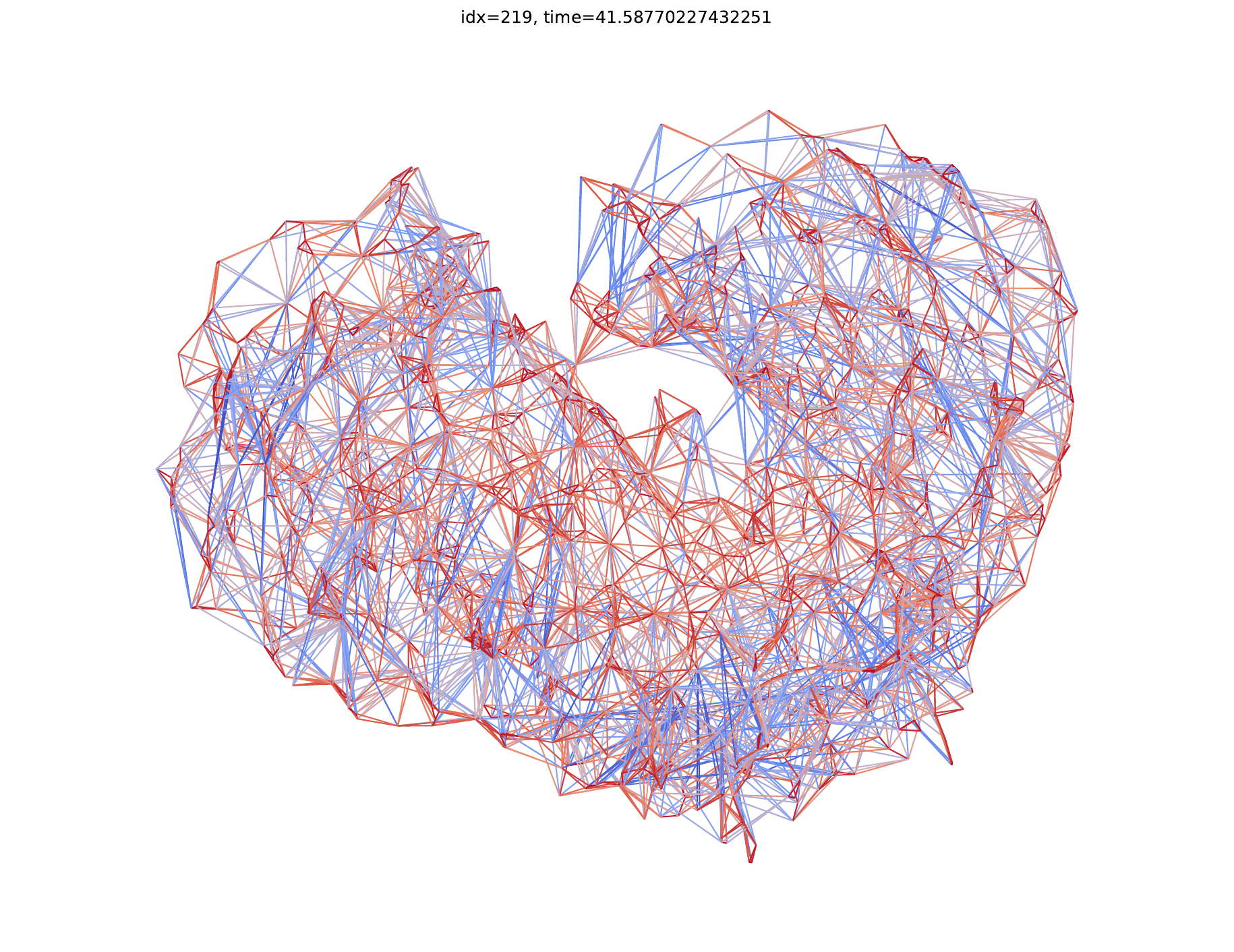} &
\imgcell{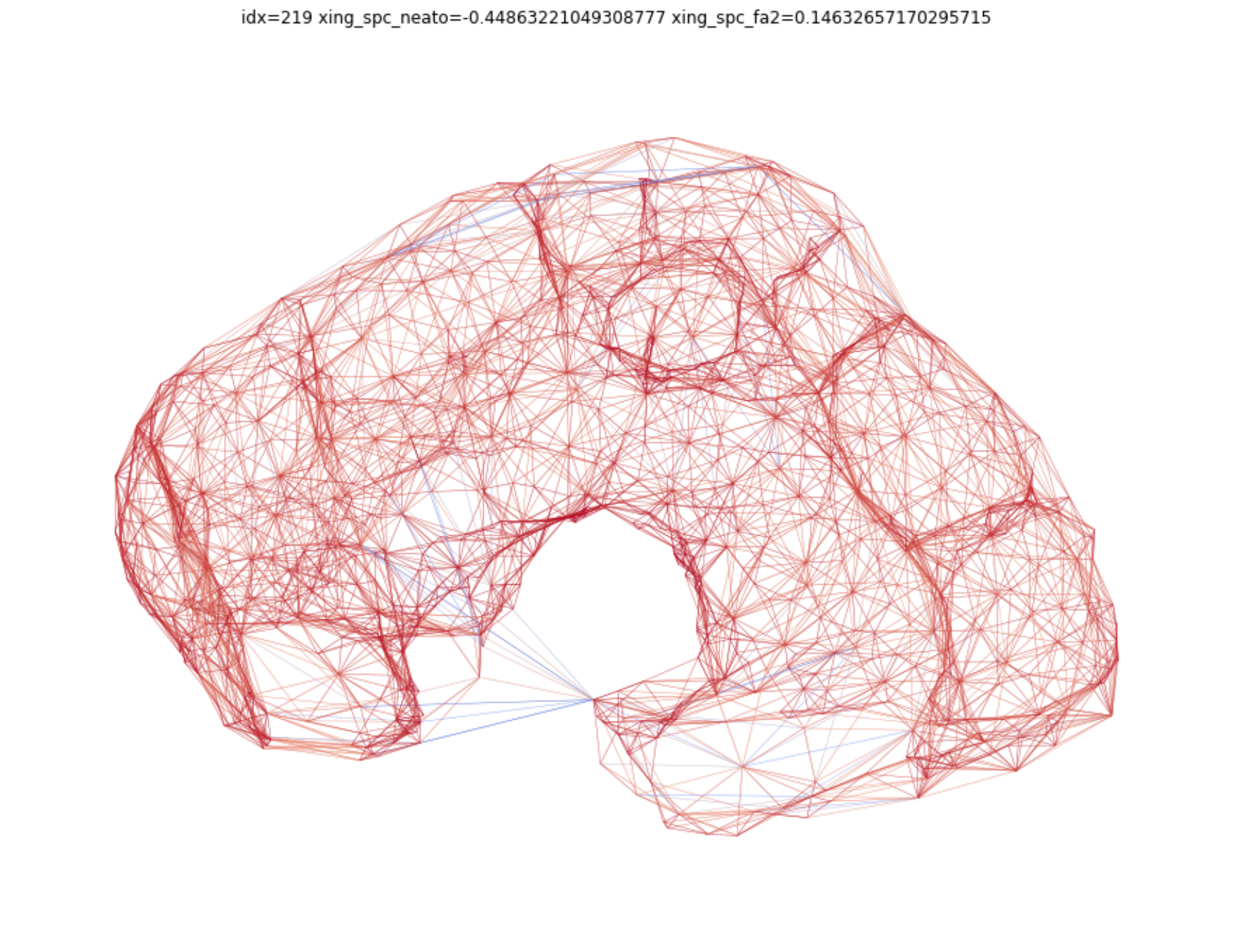} &
\imgcell{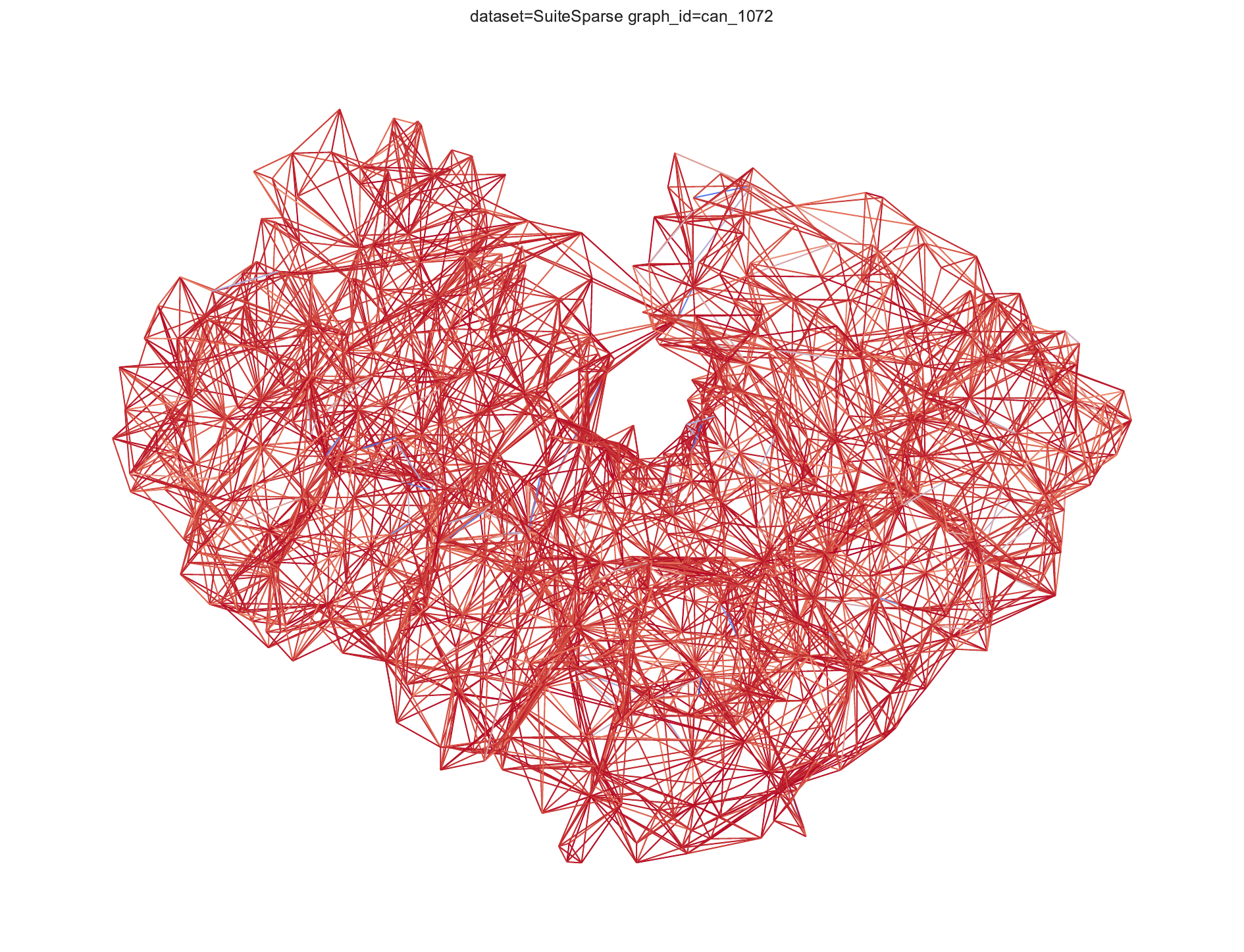} &
\imgcell{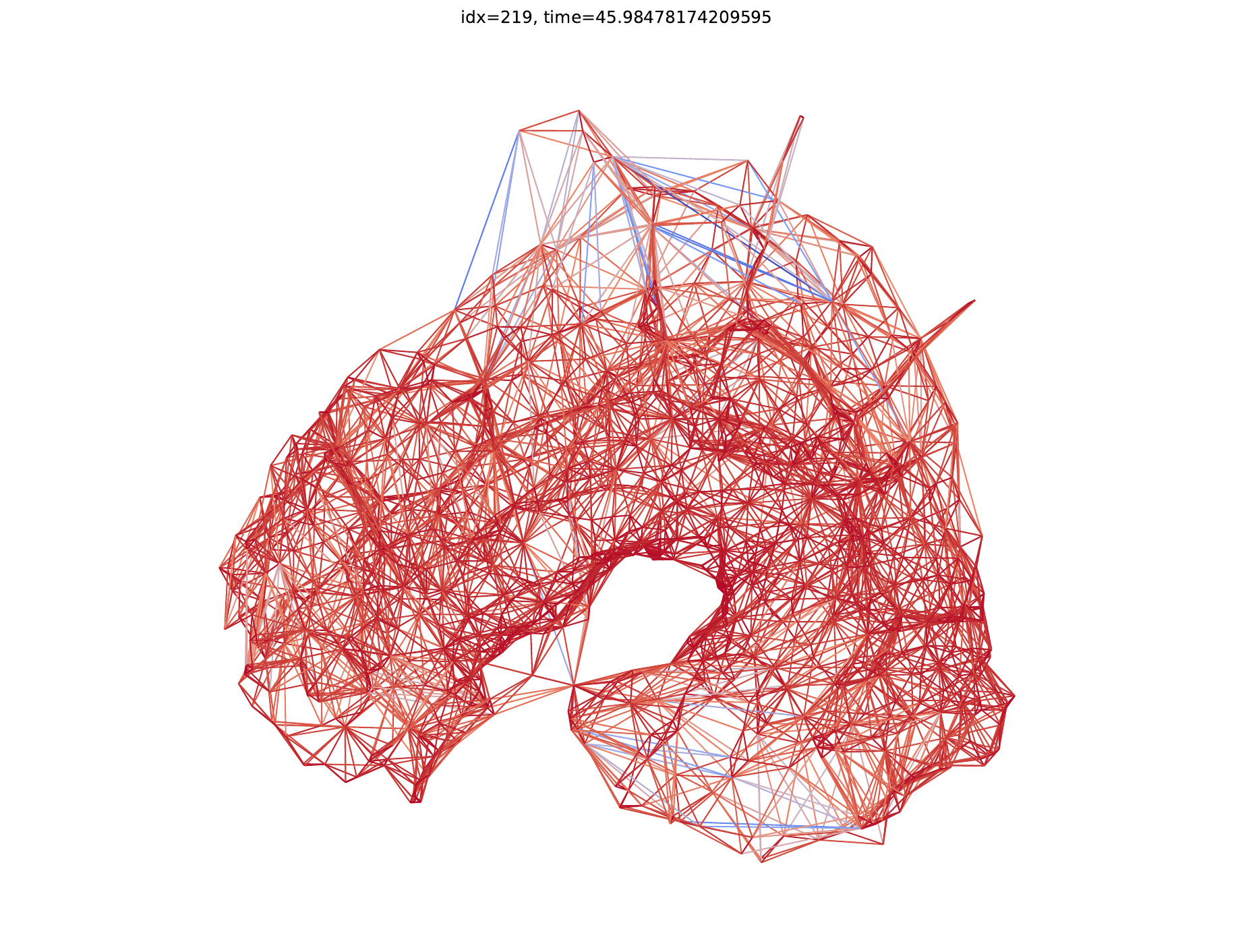} &
\imgcell{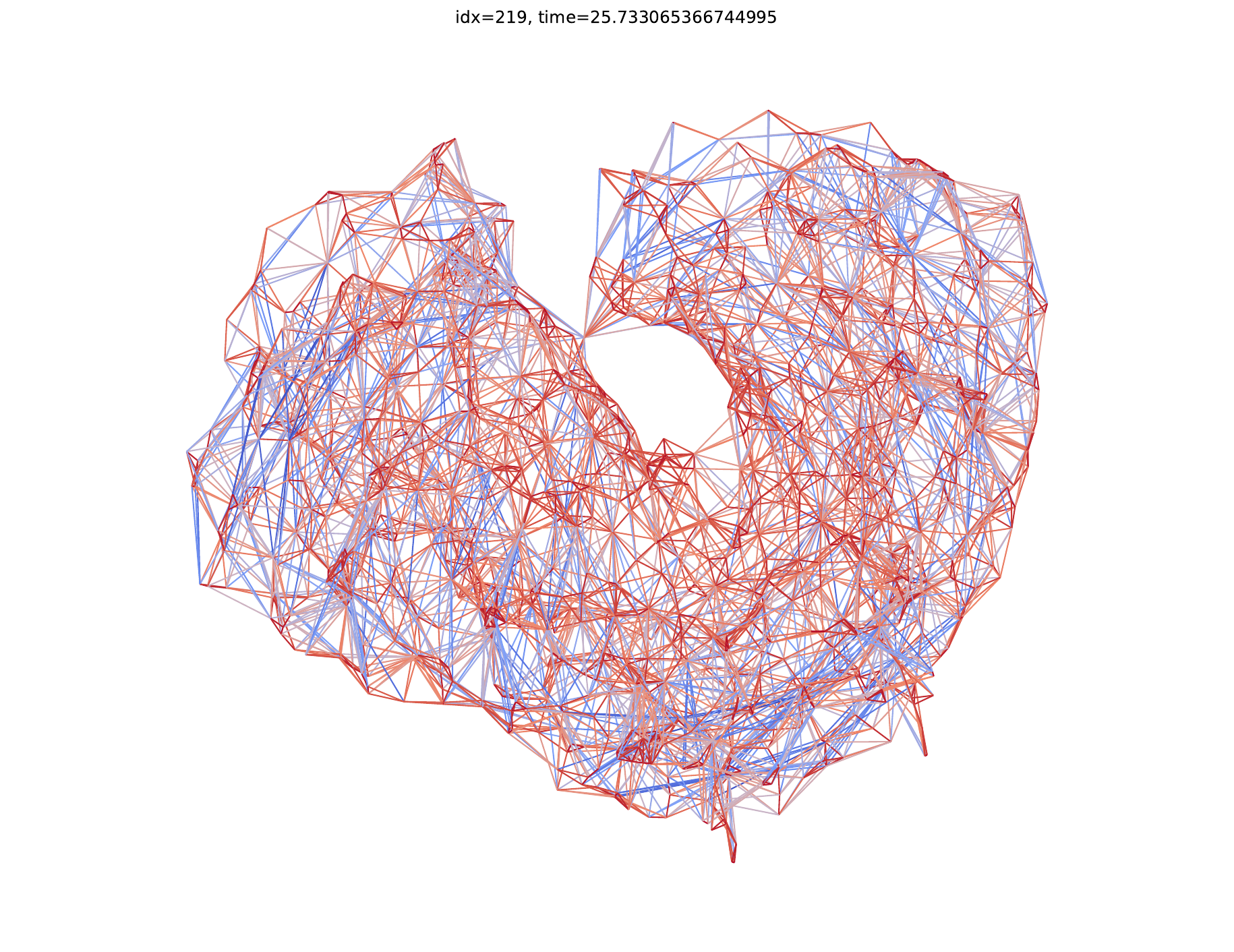} &
\imgcell{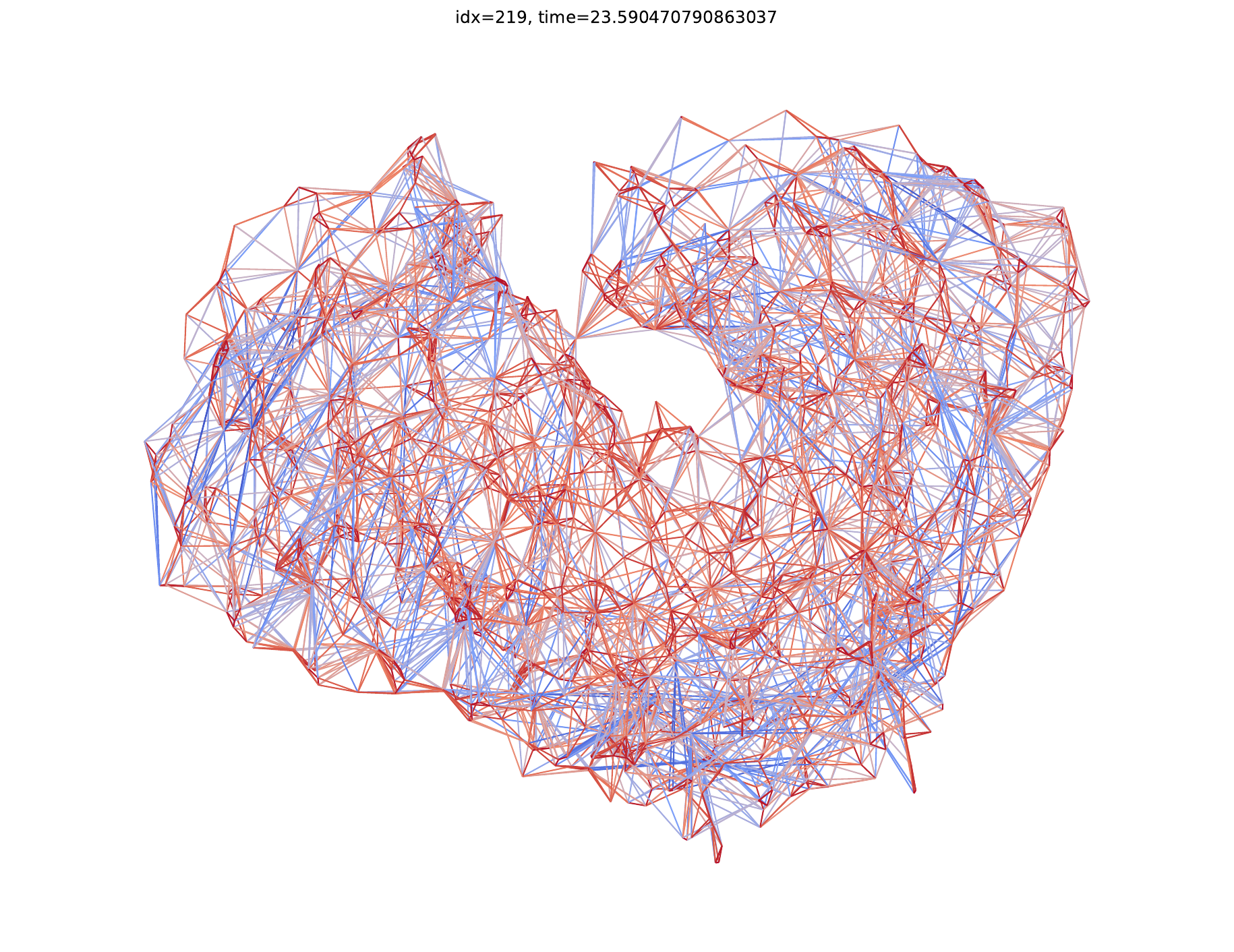} &
\imgcell{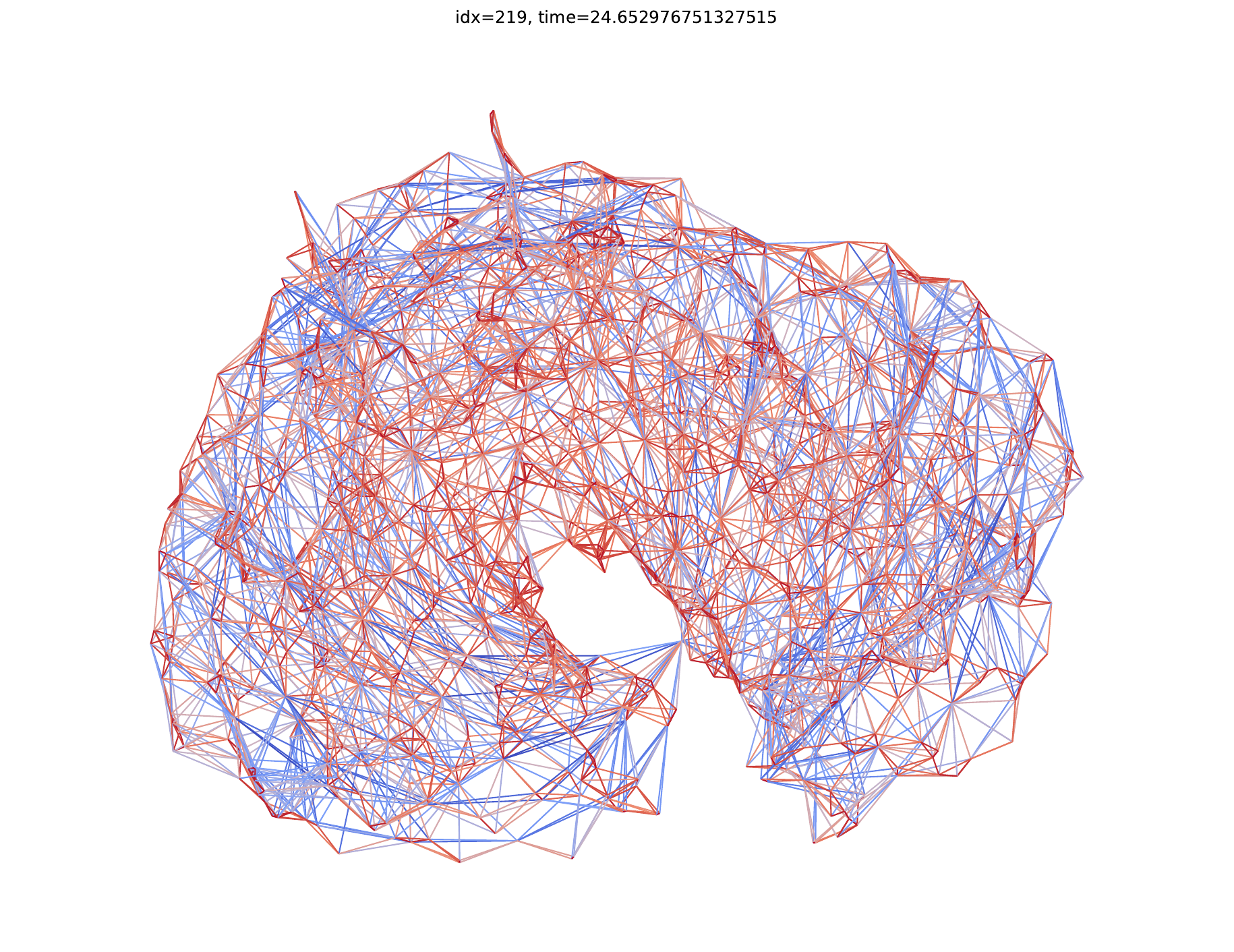} \\

&
t = 0.63s &
t = 32.28s &
t = 100.11s &
t = 4.39s &
t = 7200.00s &
t = 4.19s &
t = 4.70s &
t = 3.58s &
t = 3.23s &
t = 2.53s &
t = 3.39s &
t = 3.45s \\

\makecell{\bfseries cavity09\\N = 1182\\M = 17447} &
\imgcell{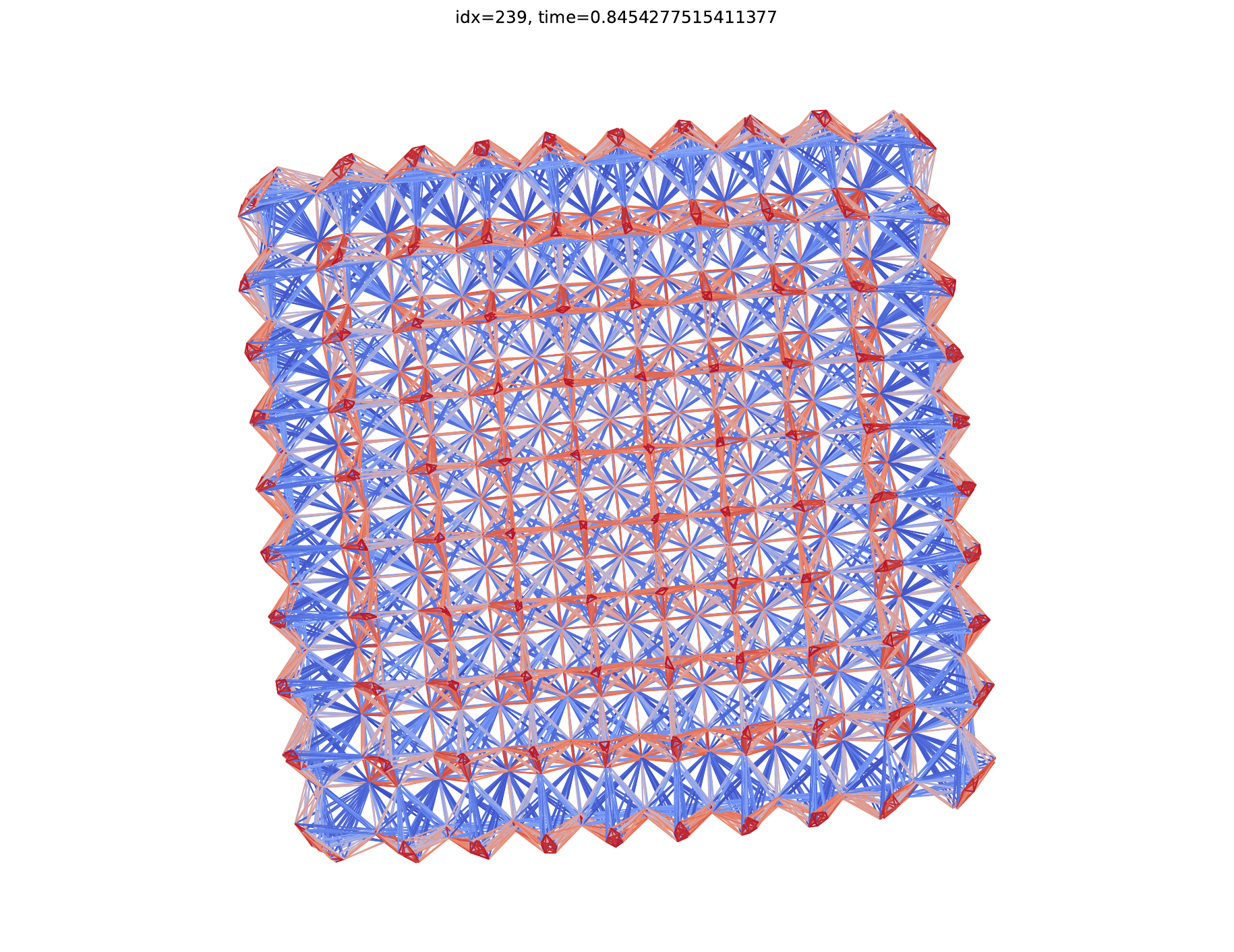} &
\imgcell{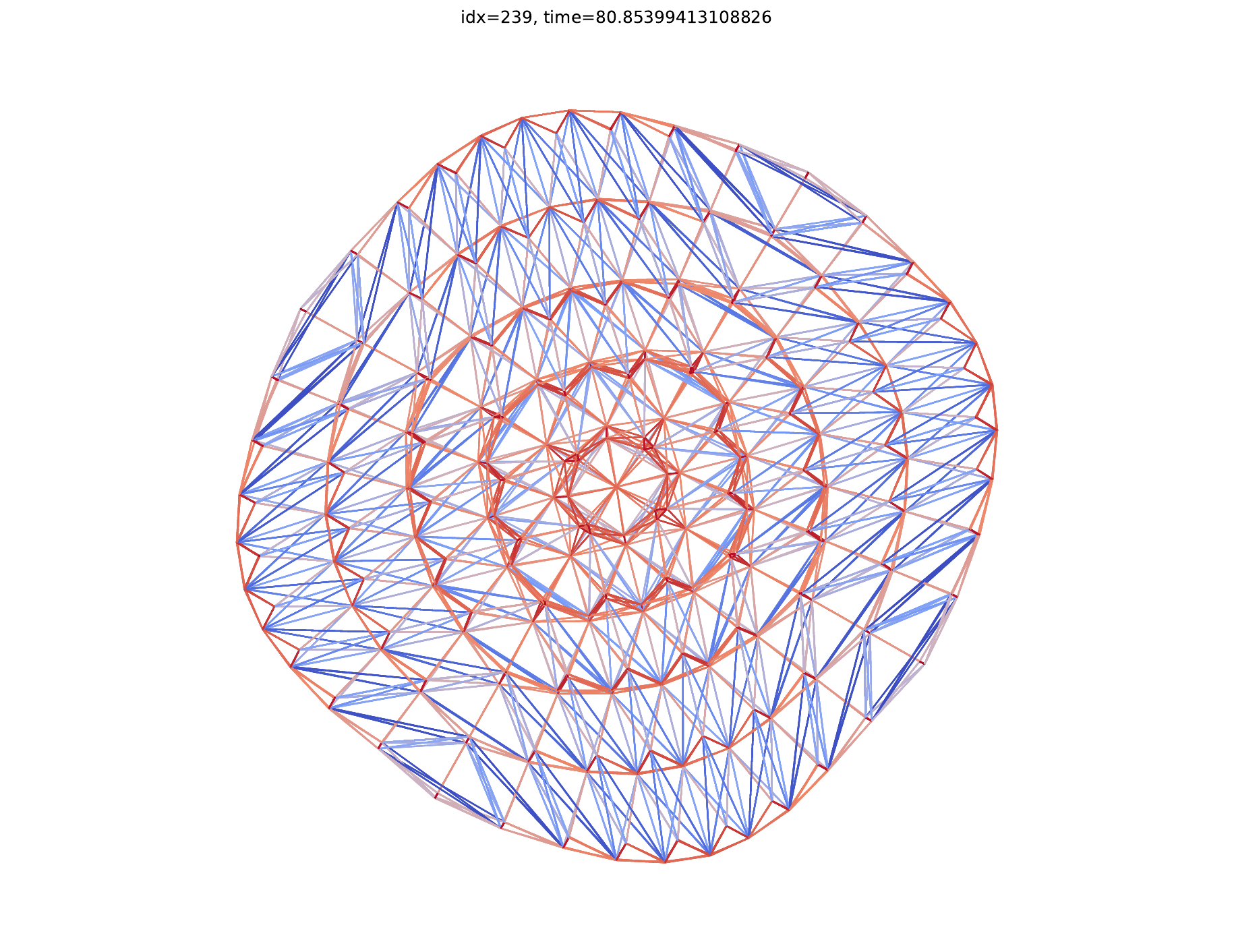} &
\imgcell{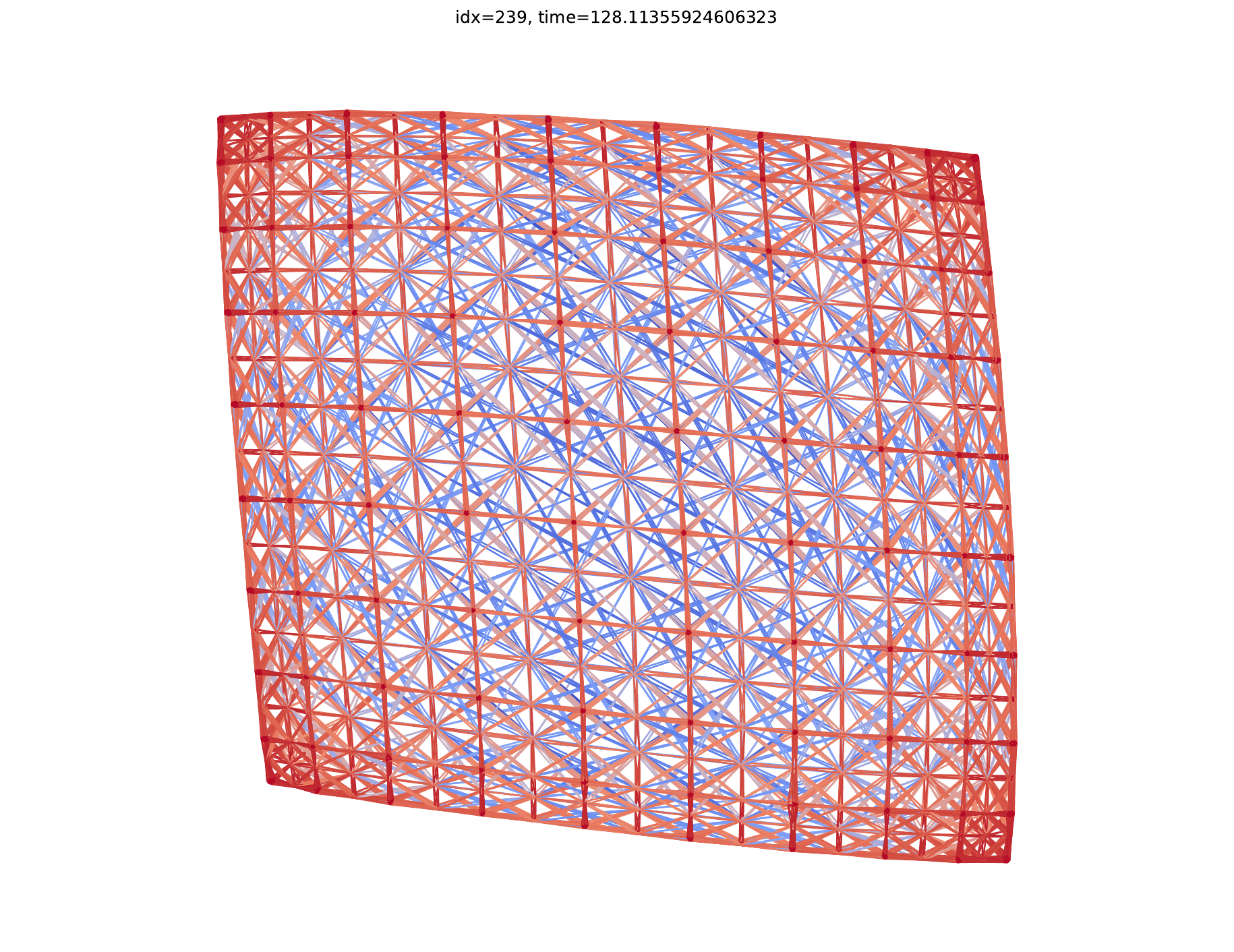} &
\imgcell{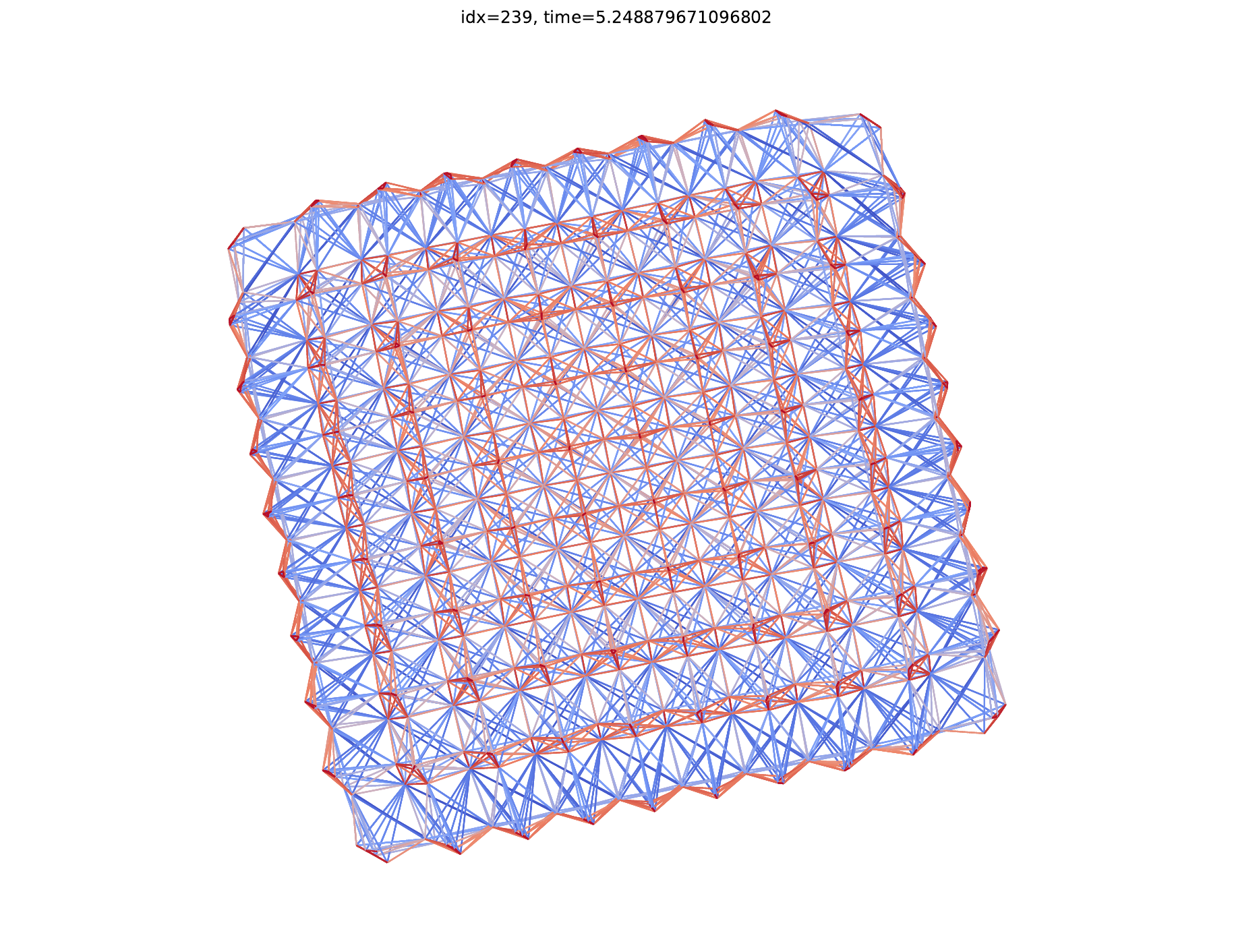} &
\imgcell{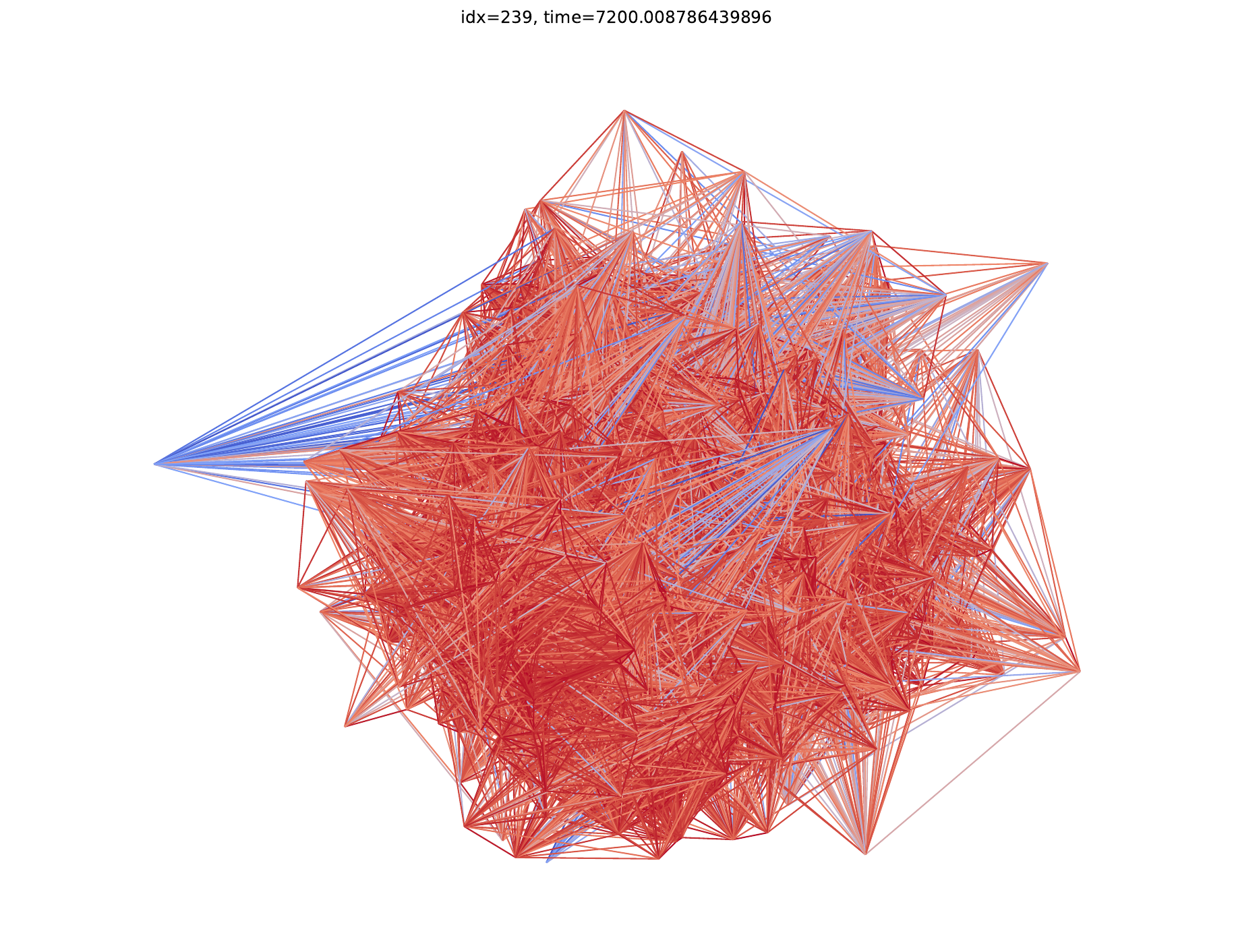} &
\imgcell{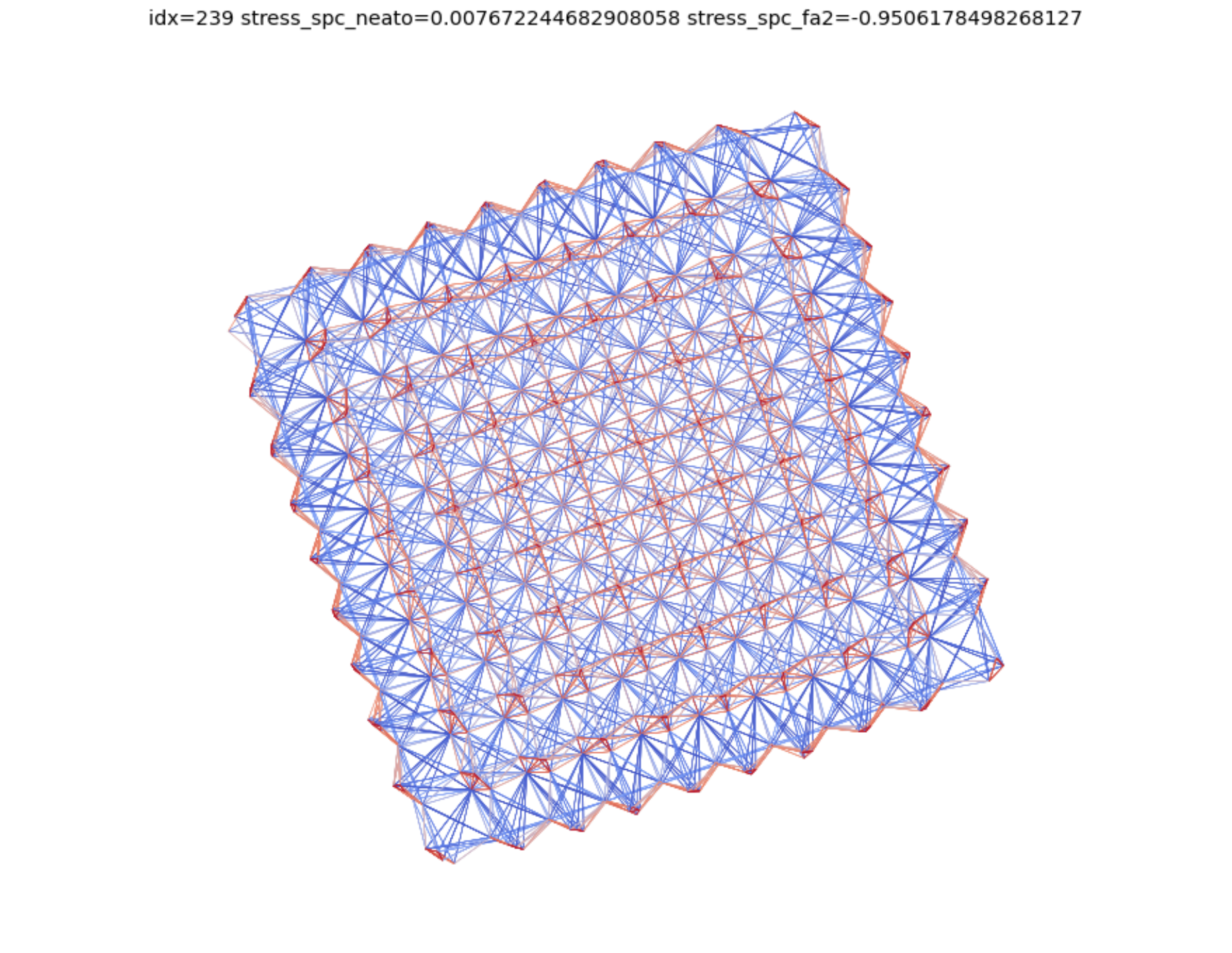} &
\imgcell{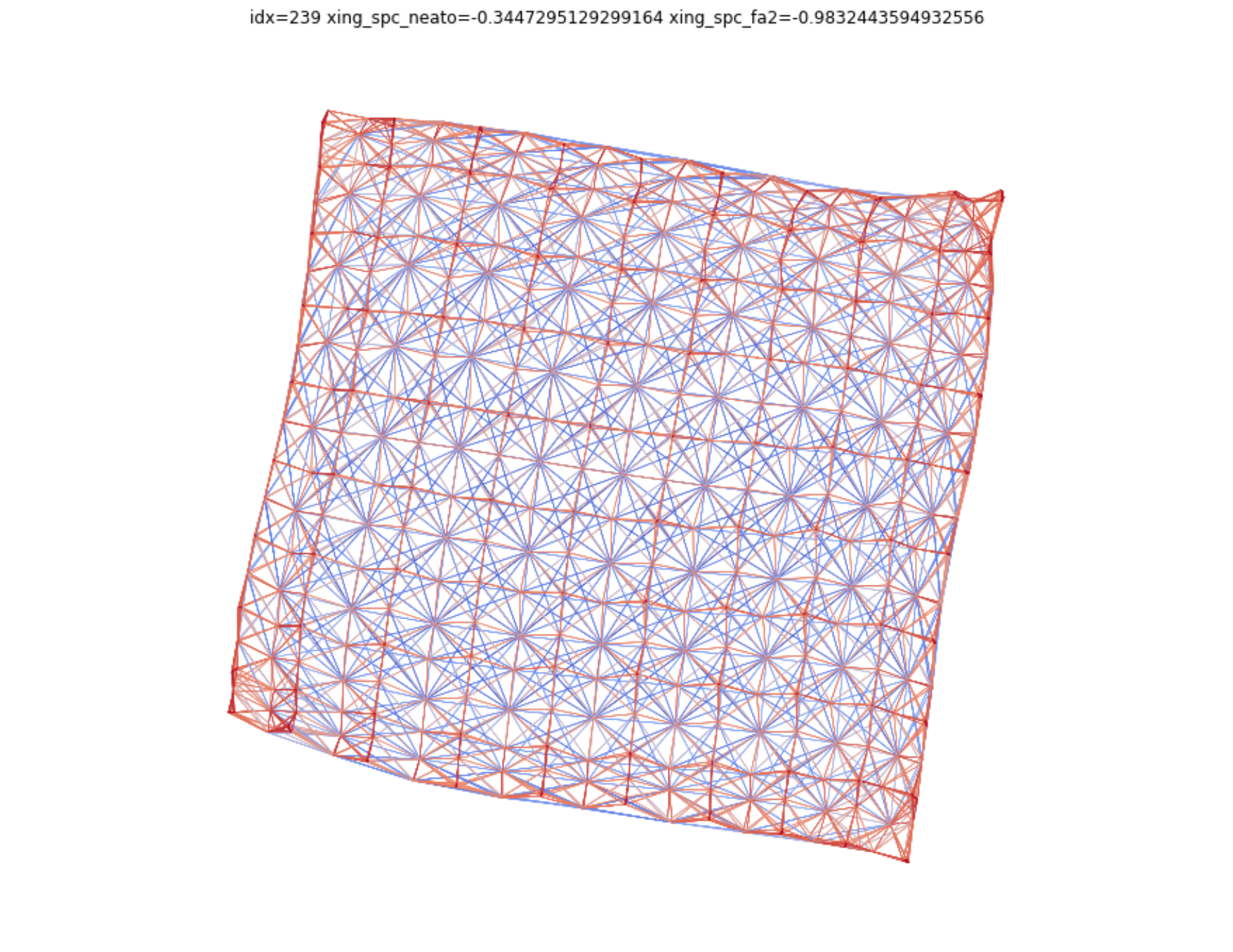} &
\imgcell{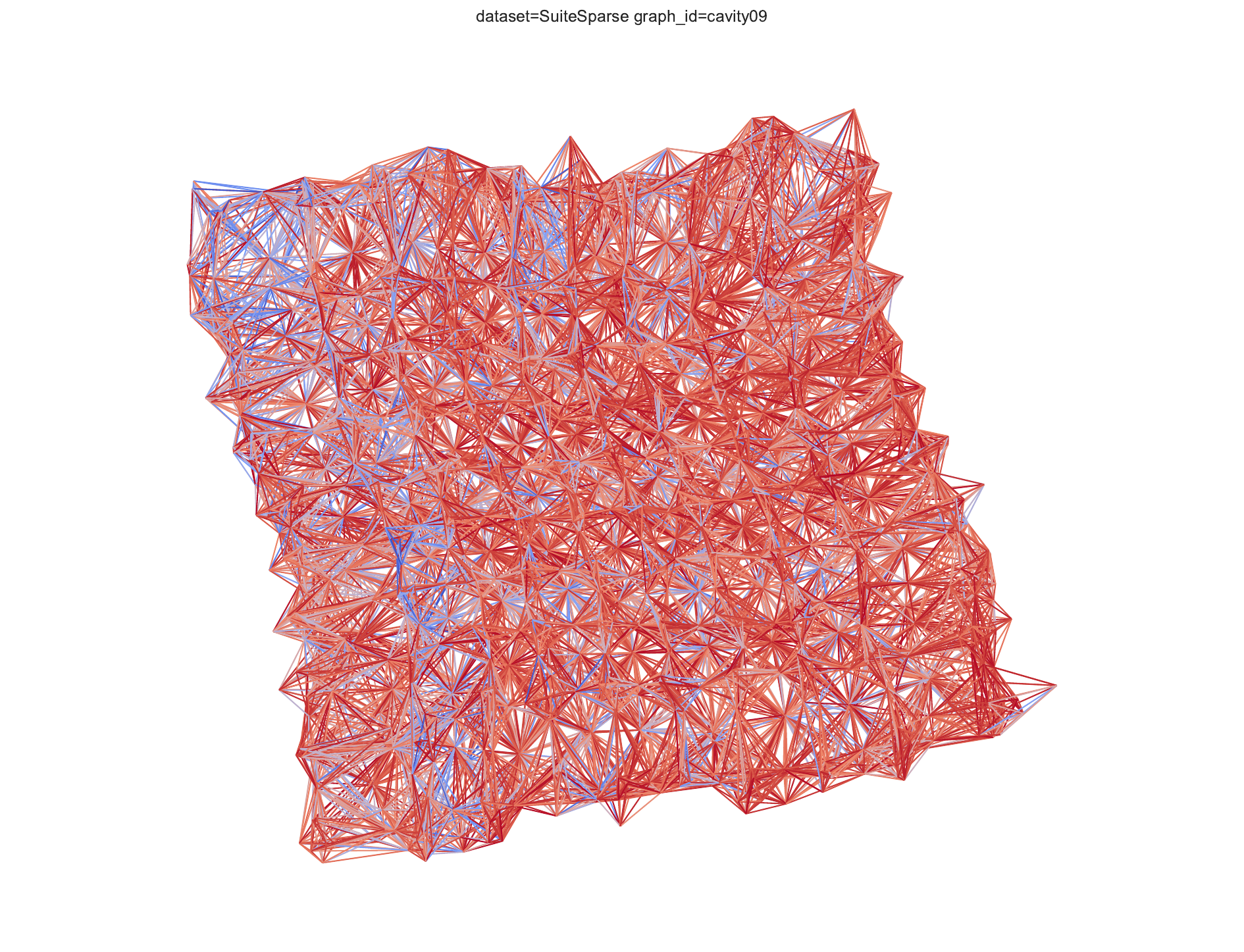} &
\imgcell{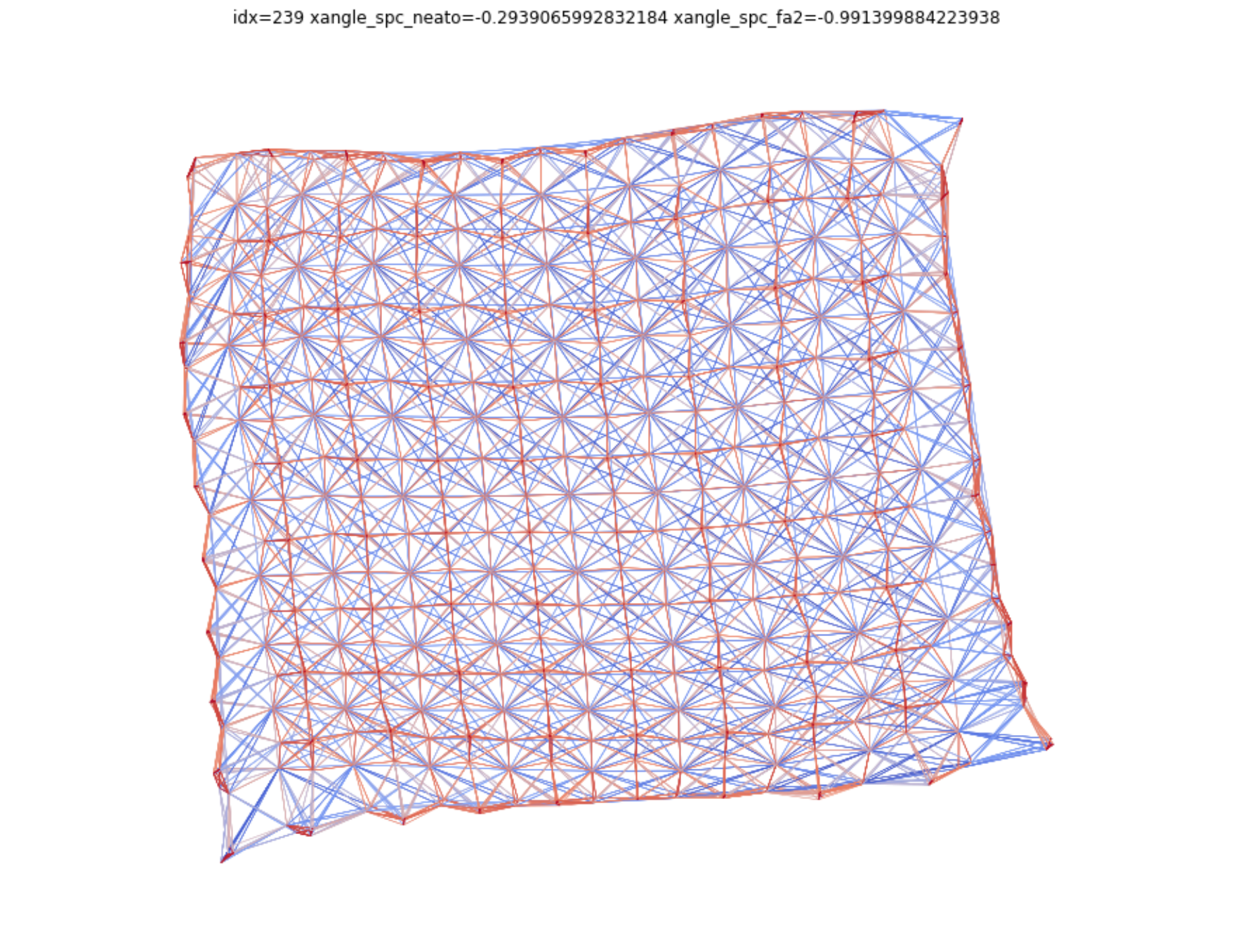} &
\imgcell{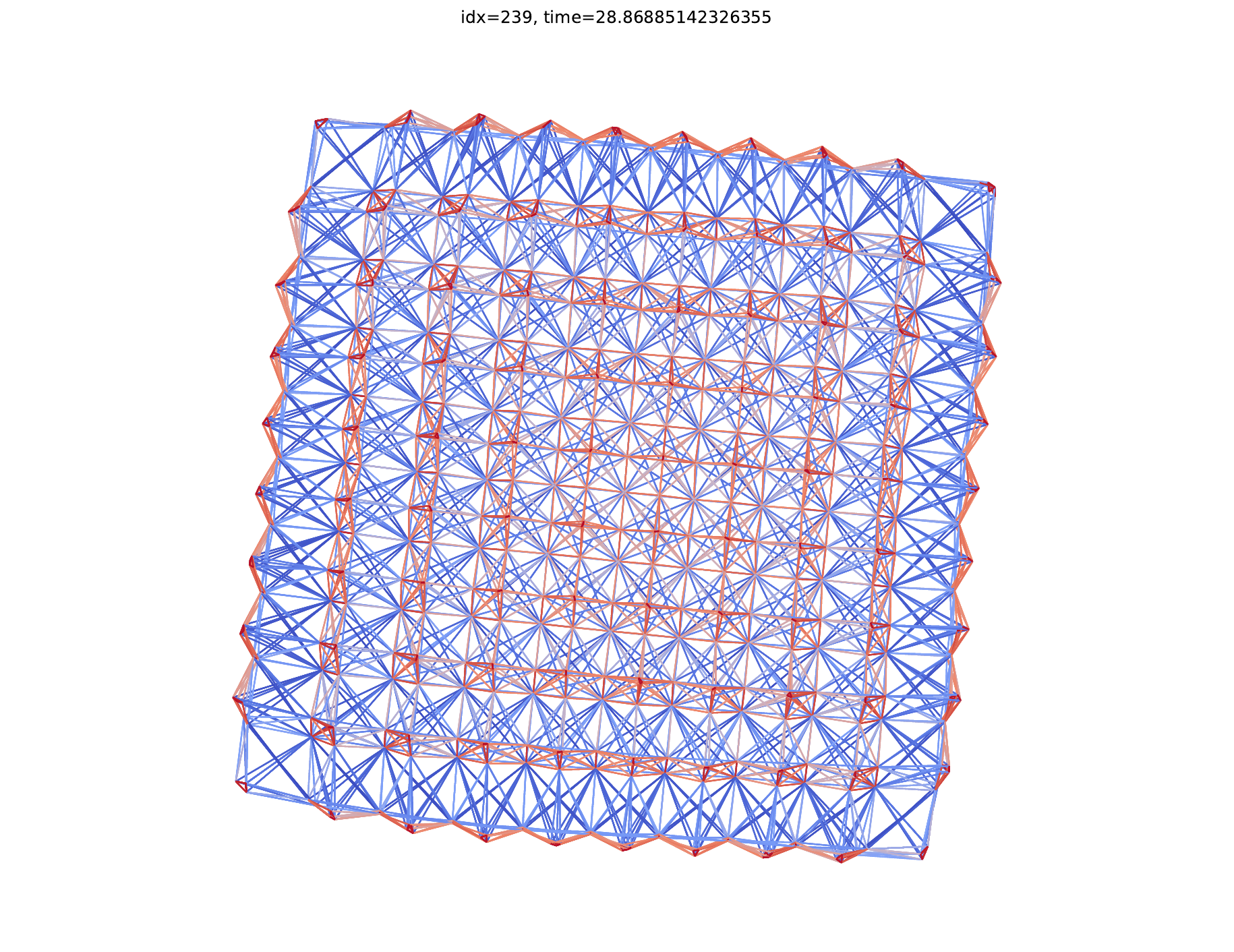} &
\imgcell{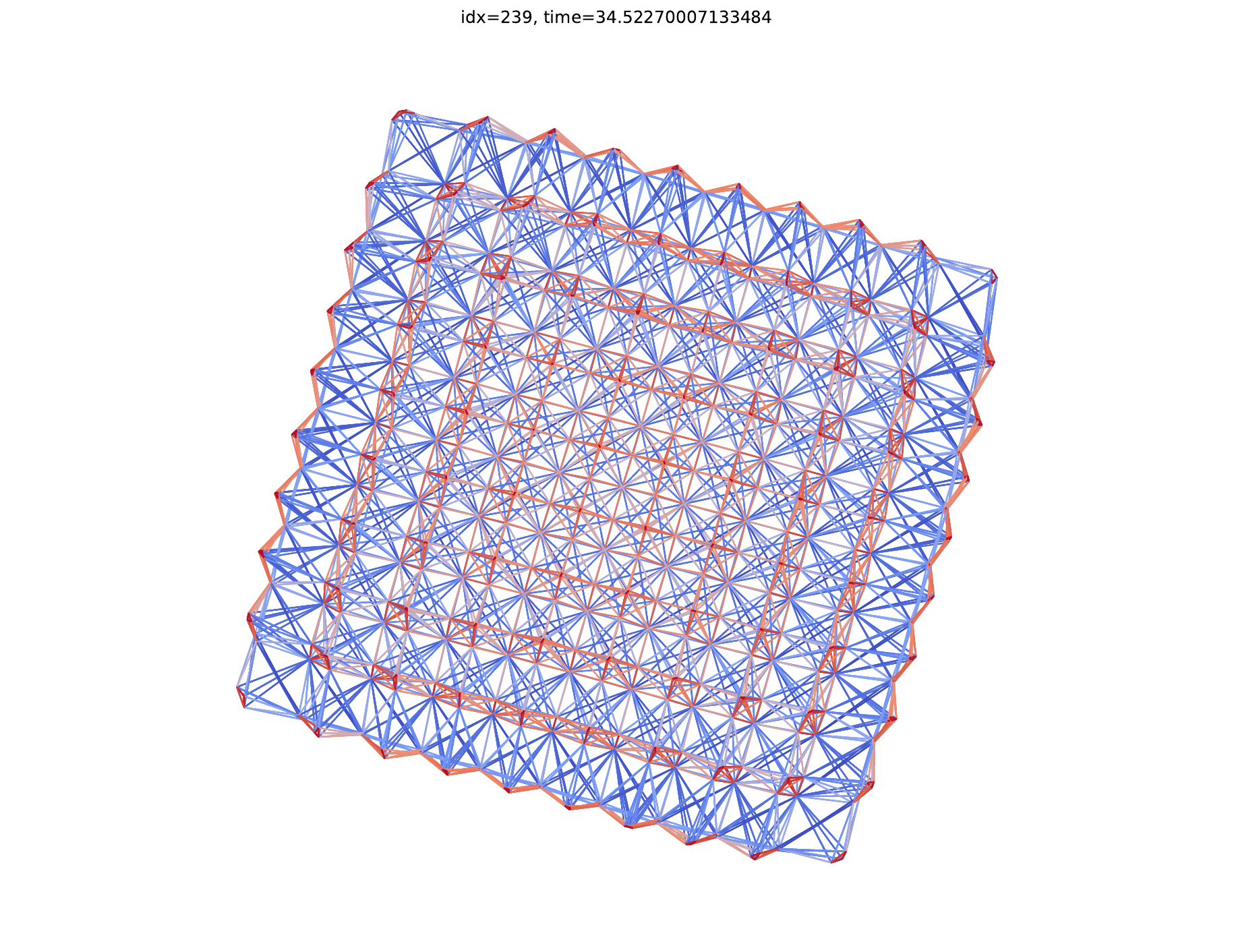} &
\imgcell{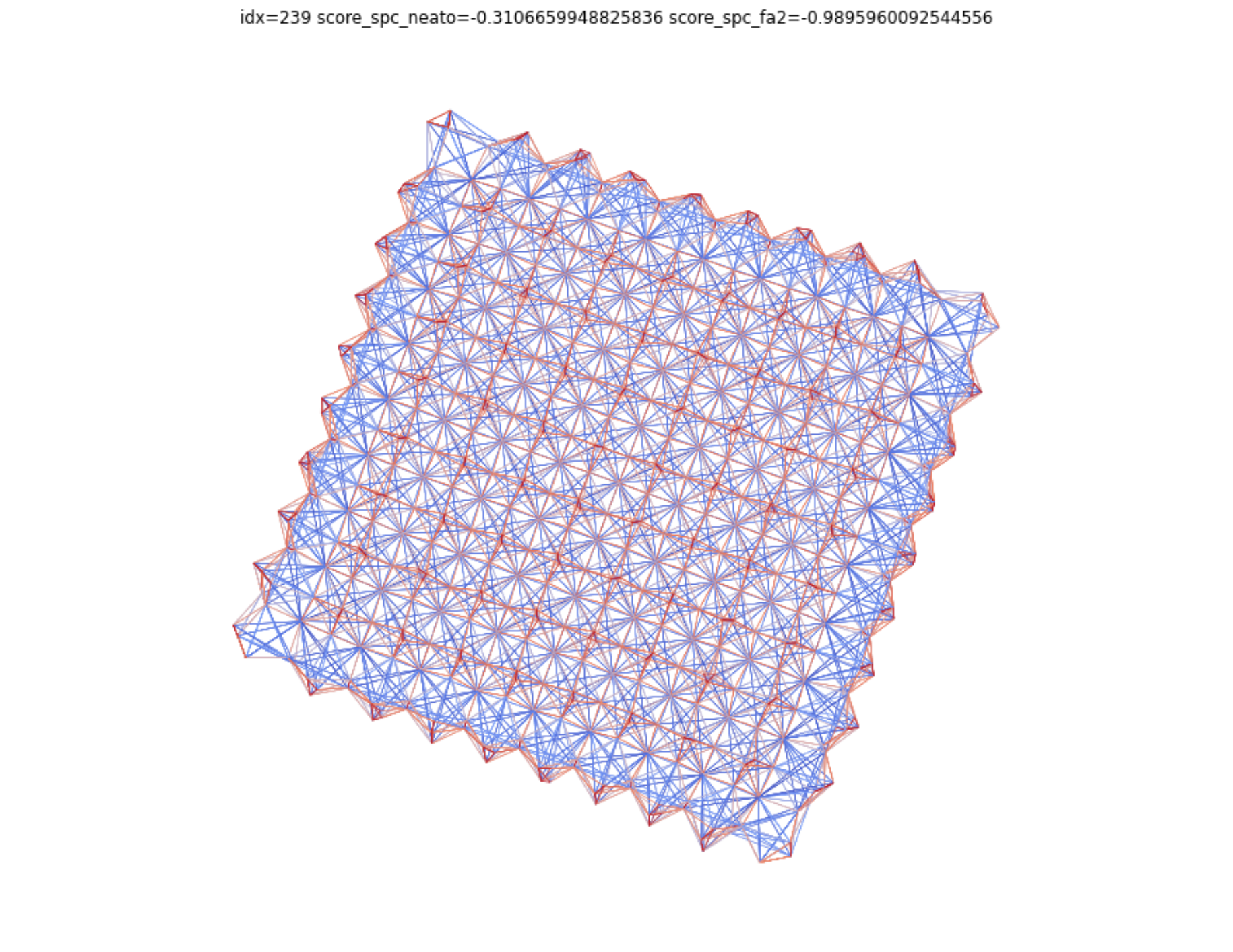} \\

&
t = 0.85s &
t = 80.85s &
t = 128.11s &
t = 5.25s &
t = 7200.00s &
t = 4.70s &
t = 4.81s &
t = 4.98s &
t = 4.58s &
t = 4.87s &
t = 5.42s &
t = 5.14s \\

\makecell{\bfseries coater1\\N = 1348\\M = 13514} &
\imgcell{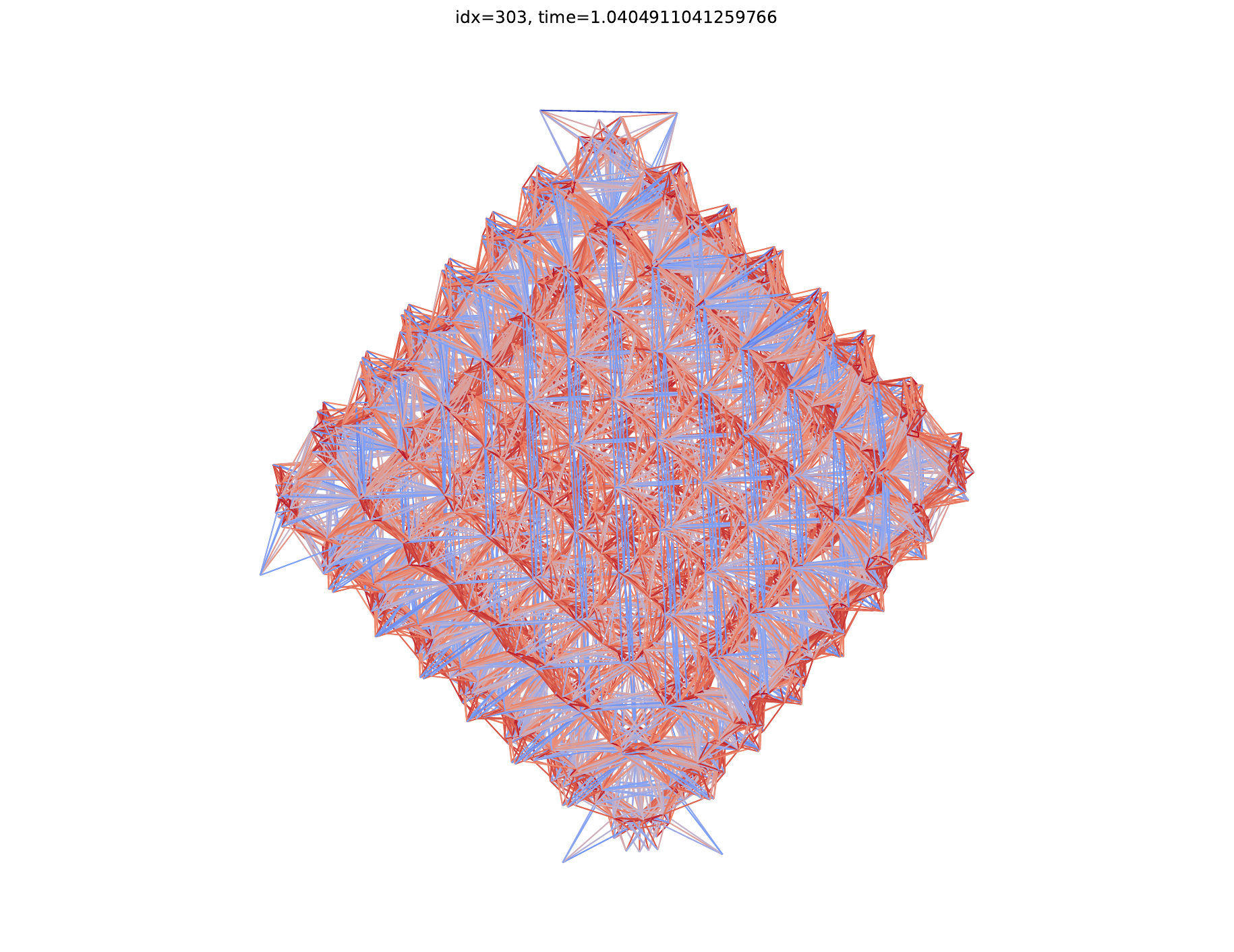} &
\imgcell{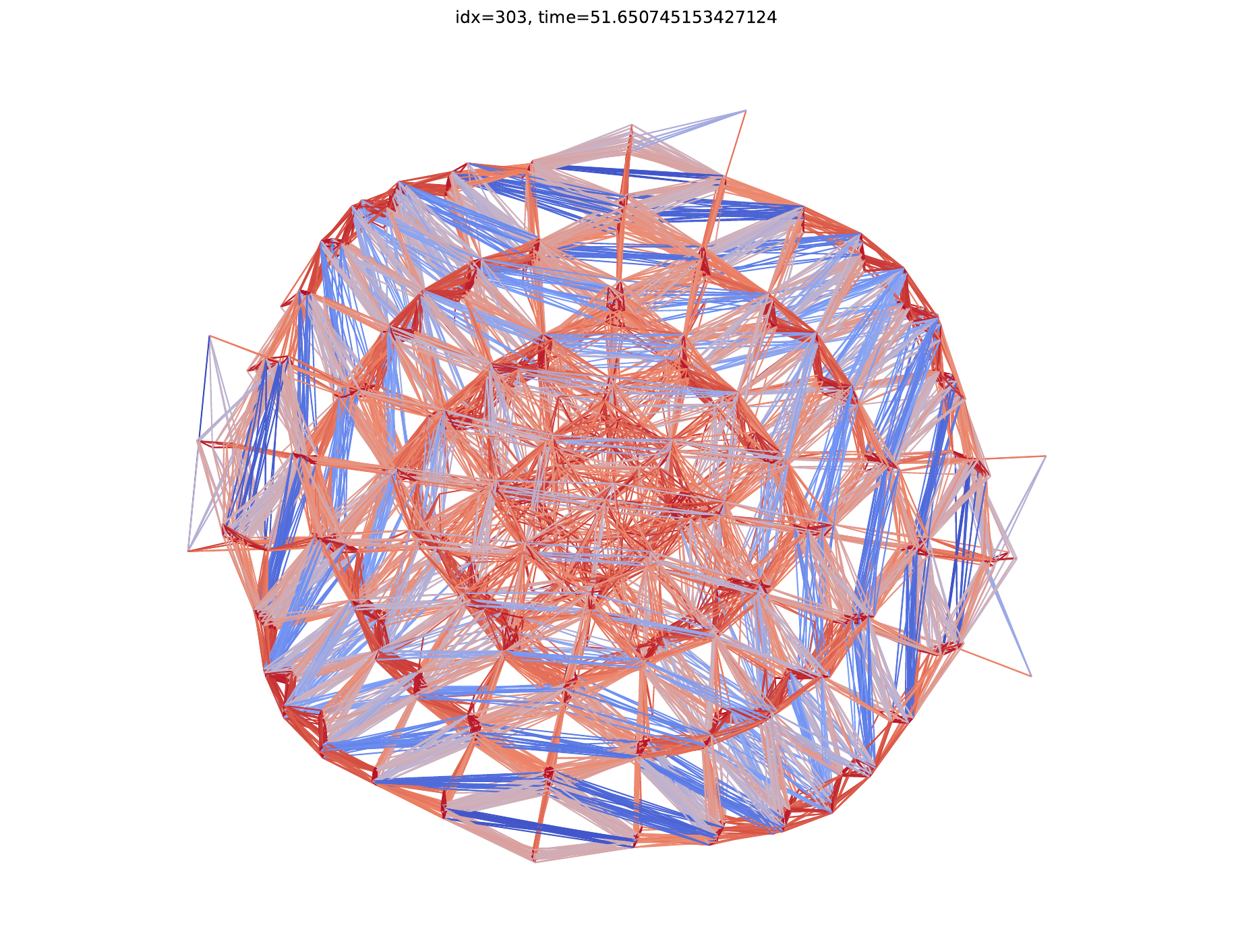} &
\imgcell{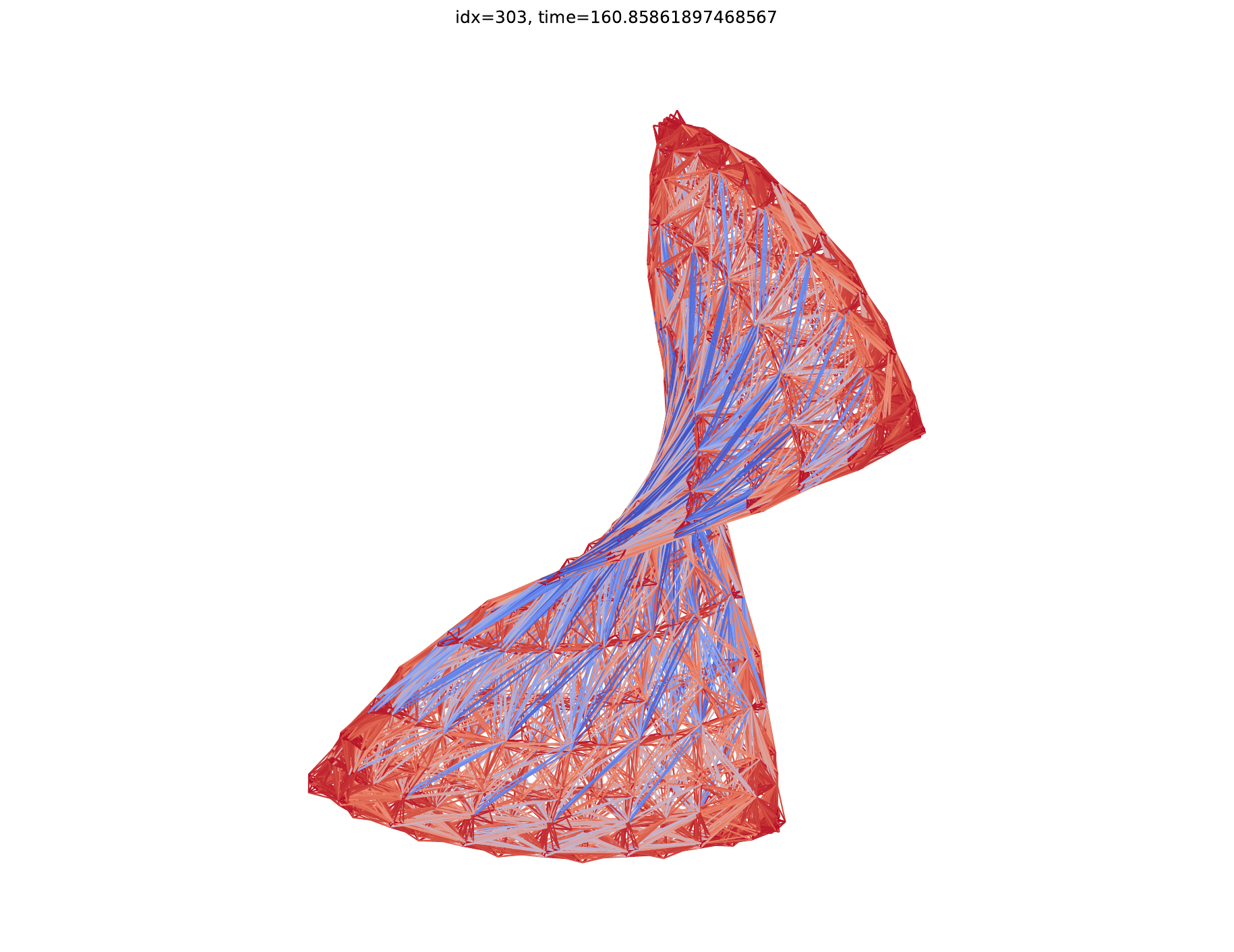} &
\imgcell{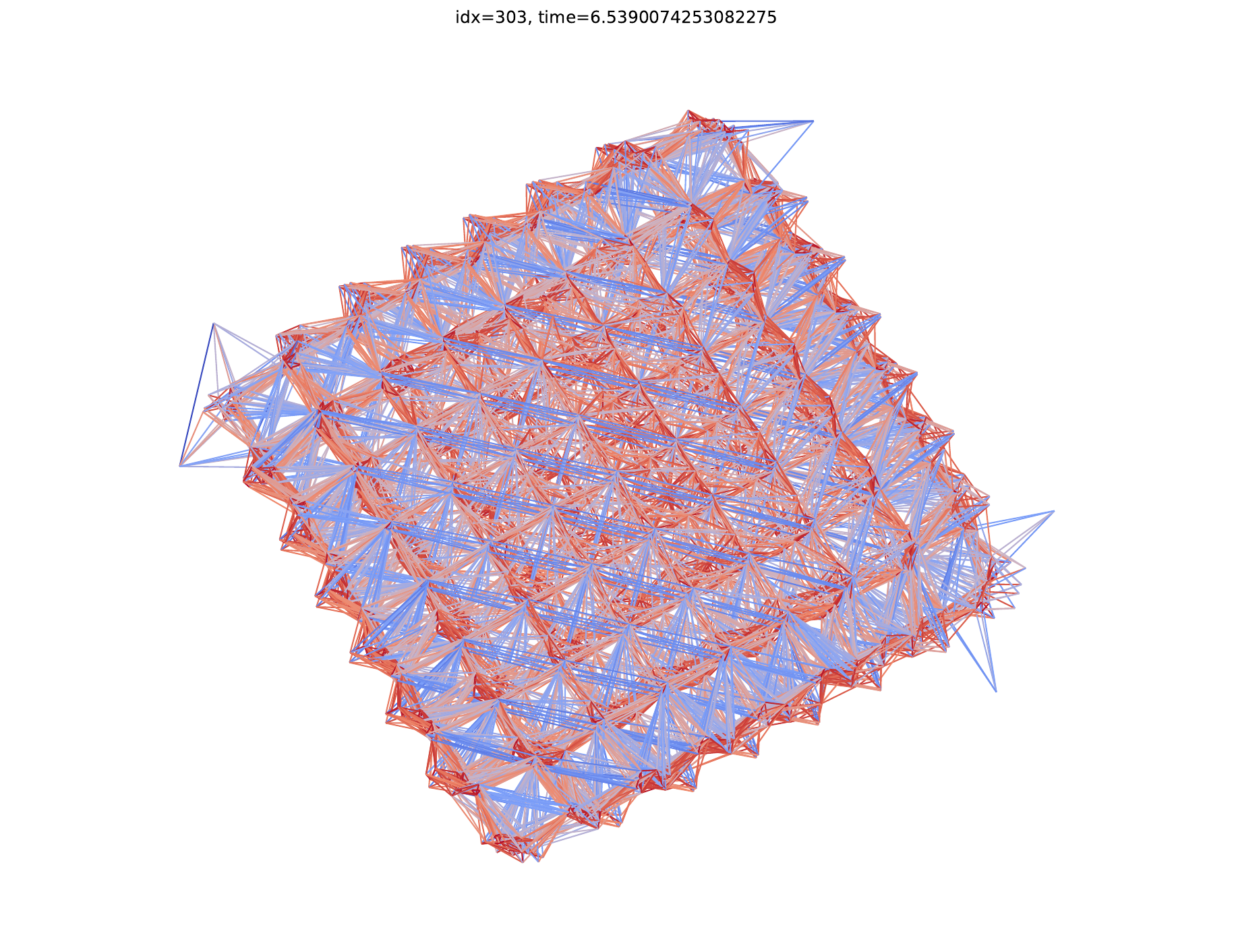} &
\imgcell{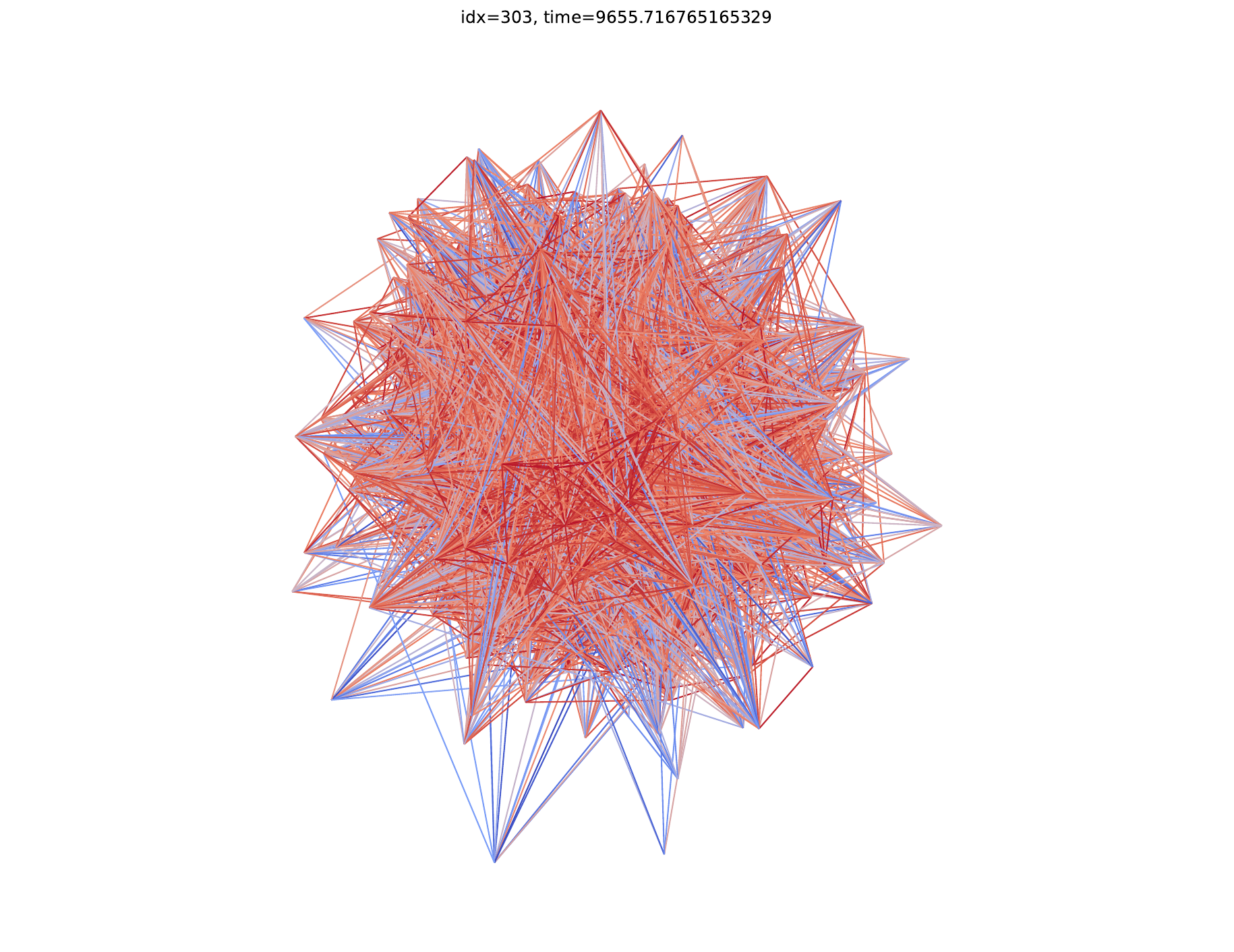} &
\imgcell{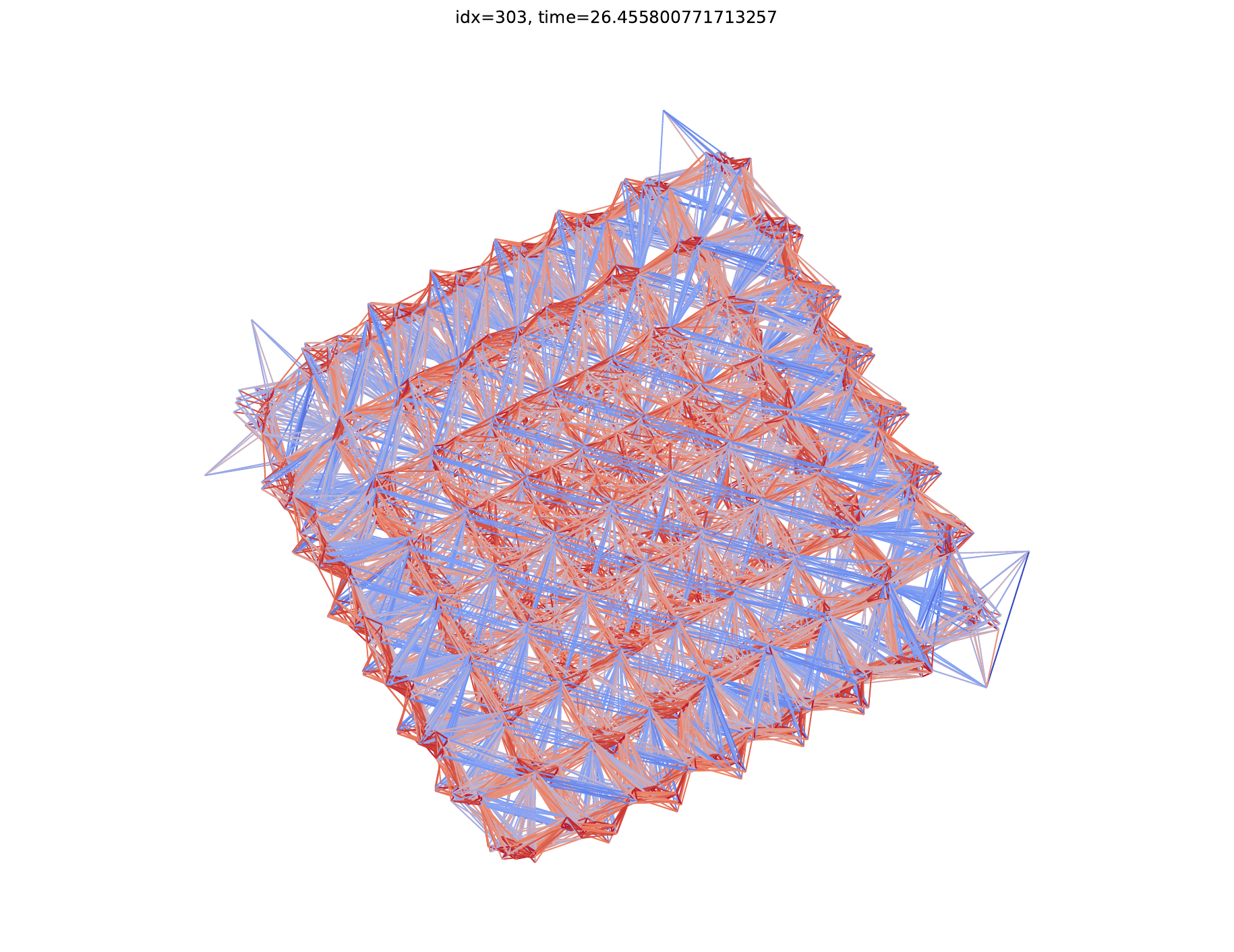} &
\imgcell{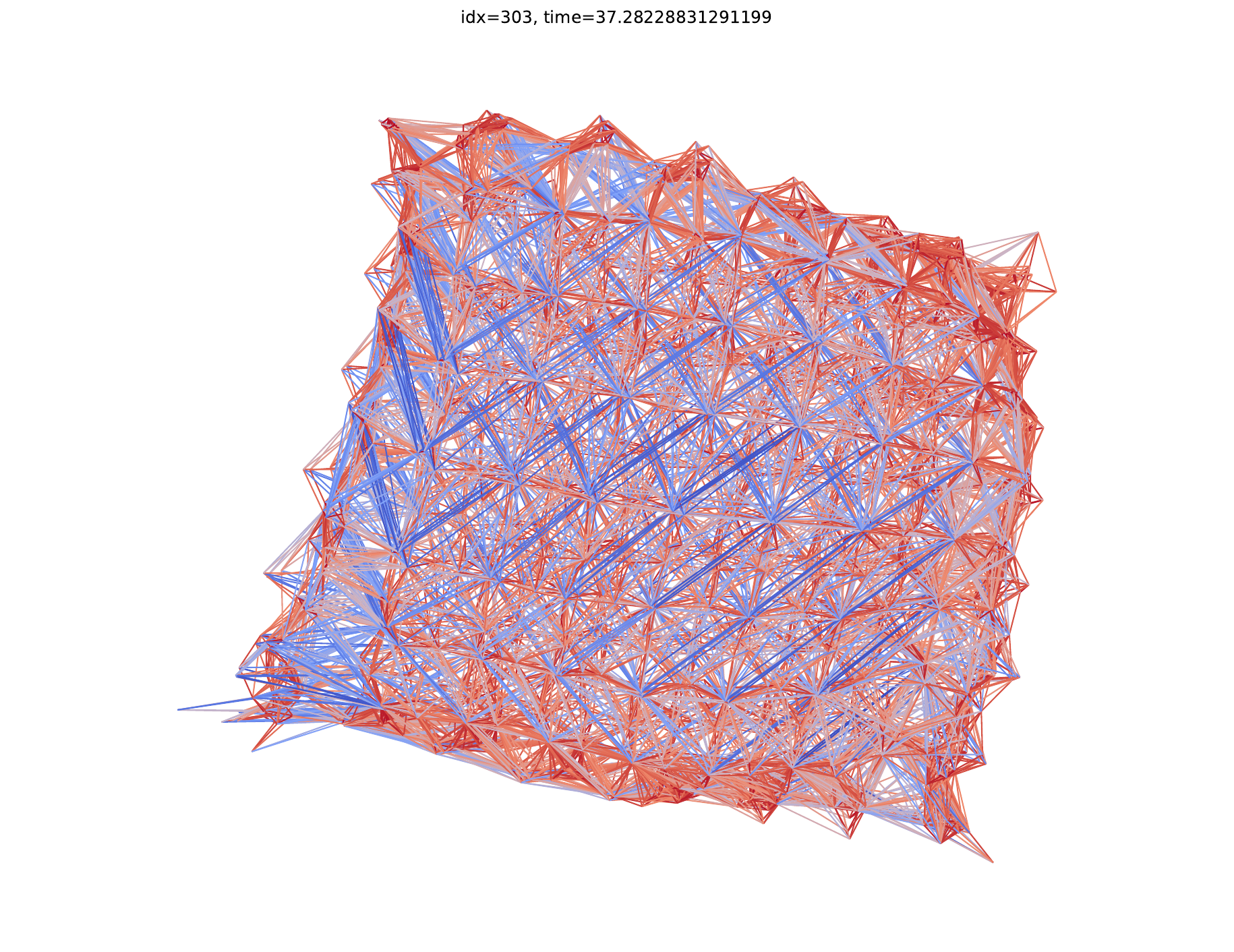} &
\imgcell{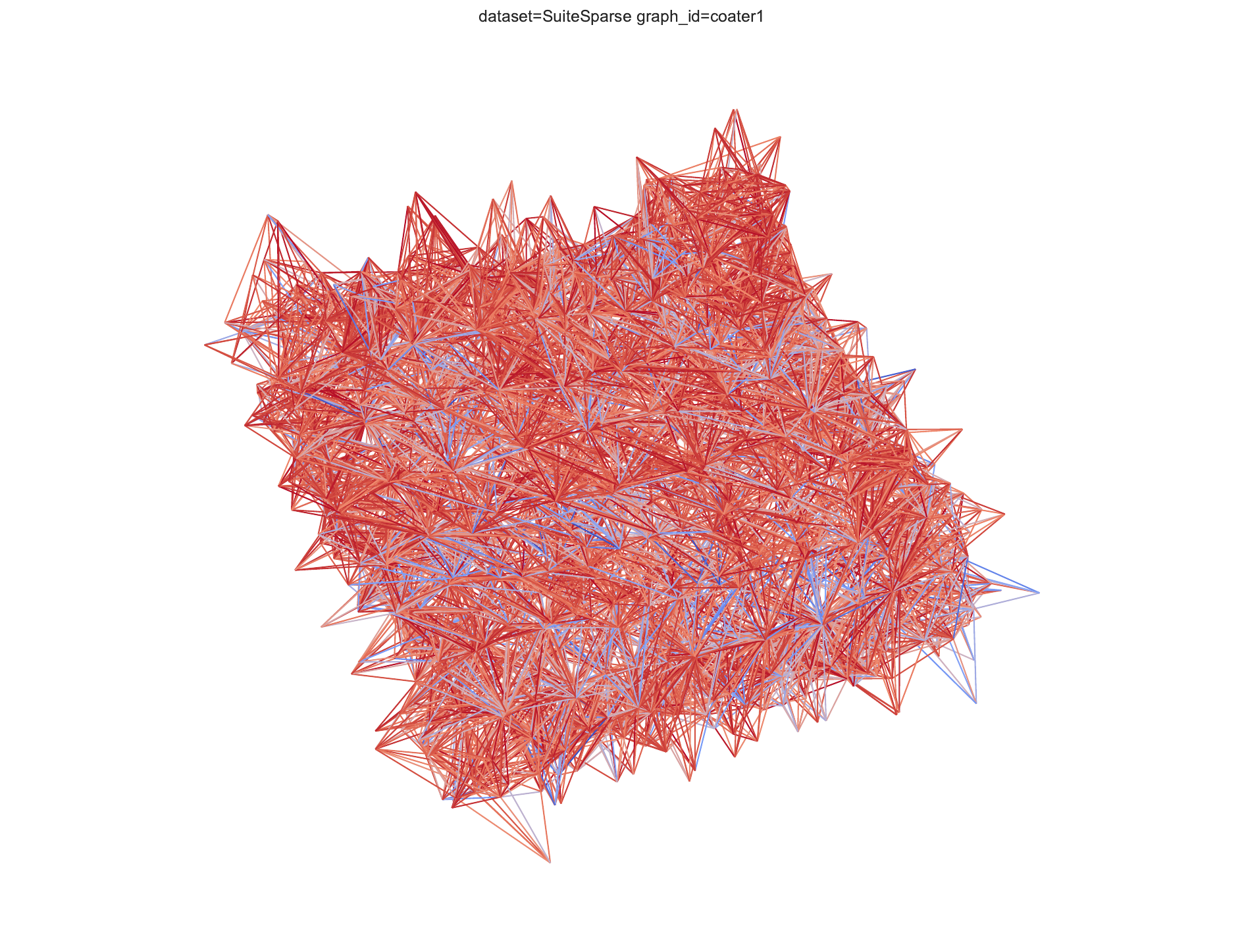} &
\imgcell{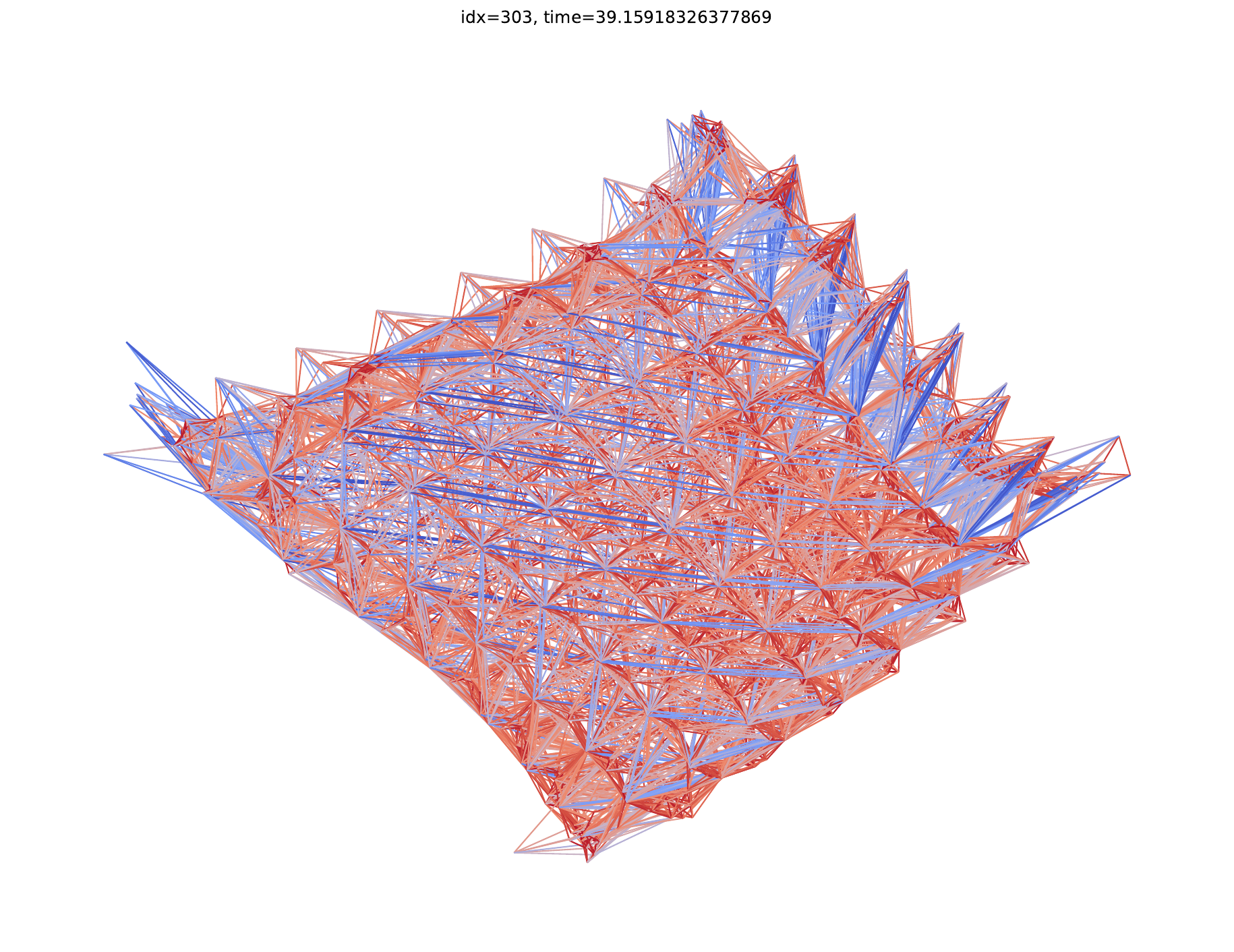} &
\imgcell{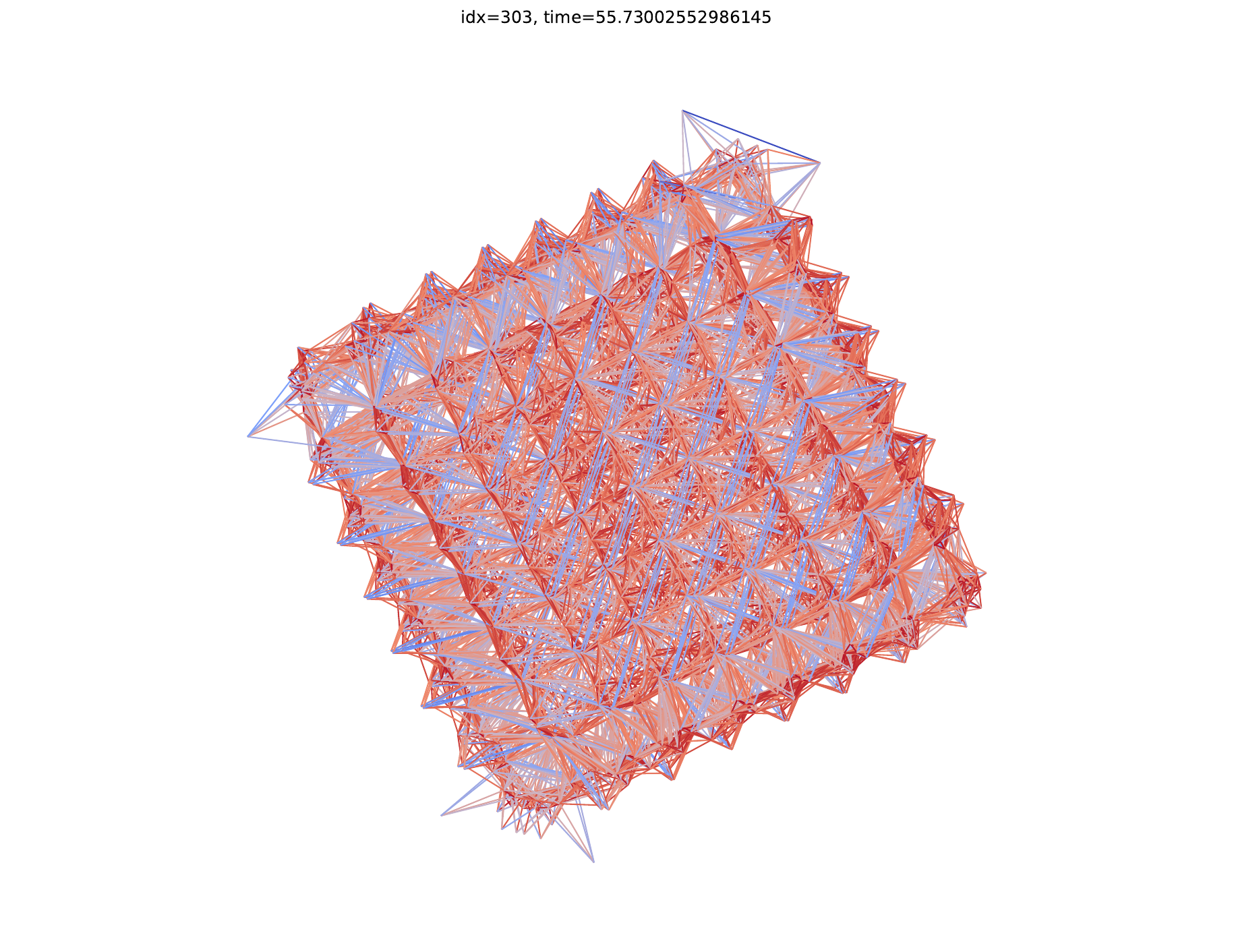} &
\imgcell{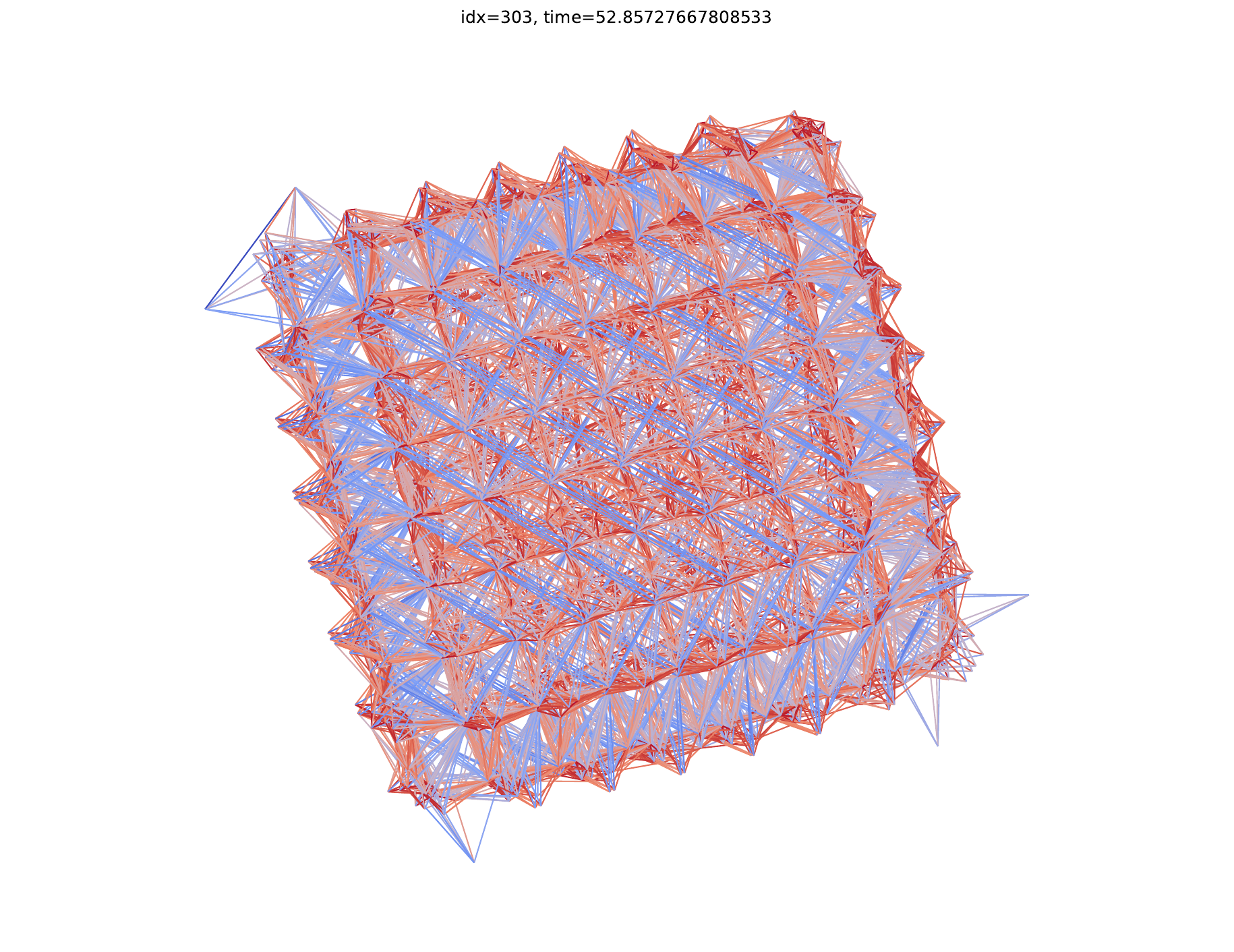} &
\imgcell{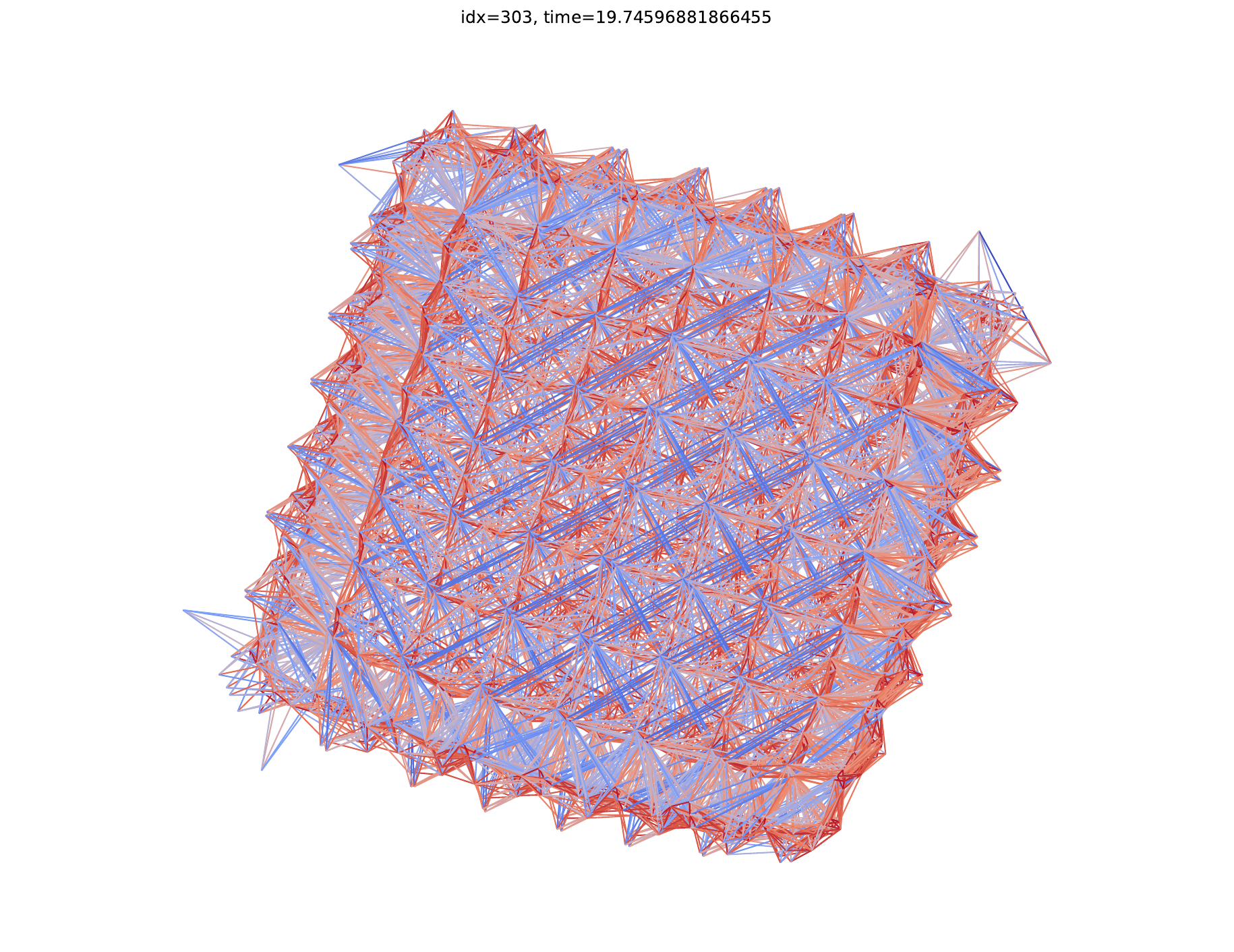} \\

&
t = 1.04s &
t = 51.65s &
t = 160.86s &
t = 6.54s &
t = 7200.00s &
t = 5.25s &
t = 5.85s &
t = 5.45s &
t = 5.60s &
t = 5.53s &
t = 6.26s &
t = 5.49s \\

\makecell{\bfseries nnc1374\\N = 1374\\M = 4576} &
\imgcell{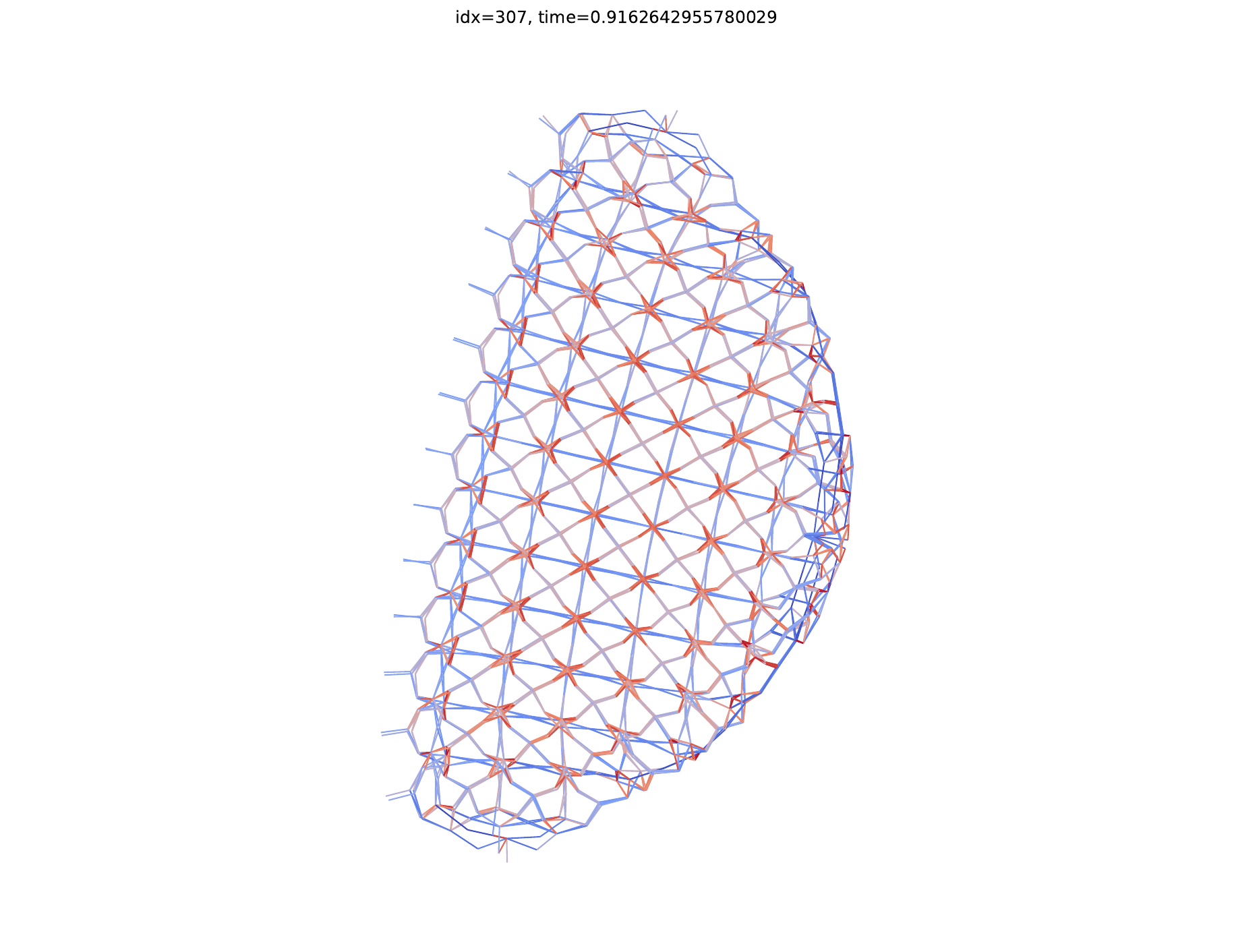} &
\imgcell{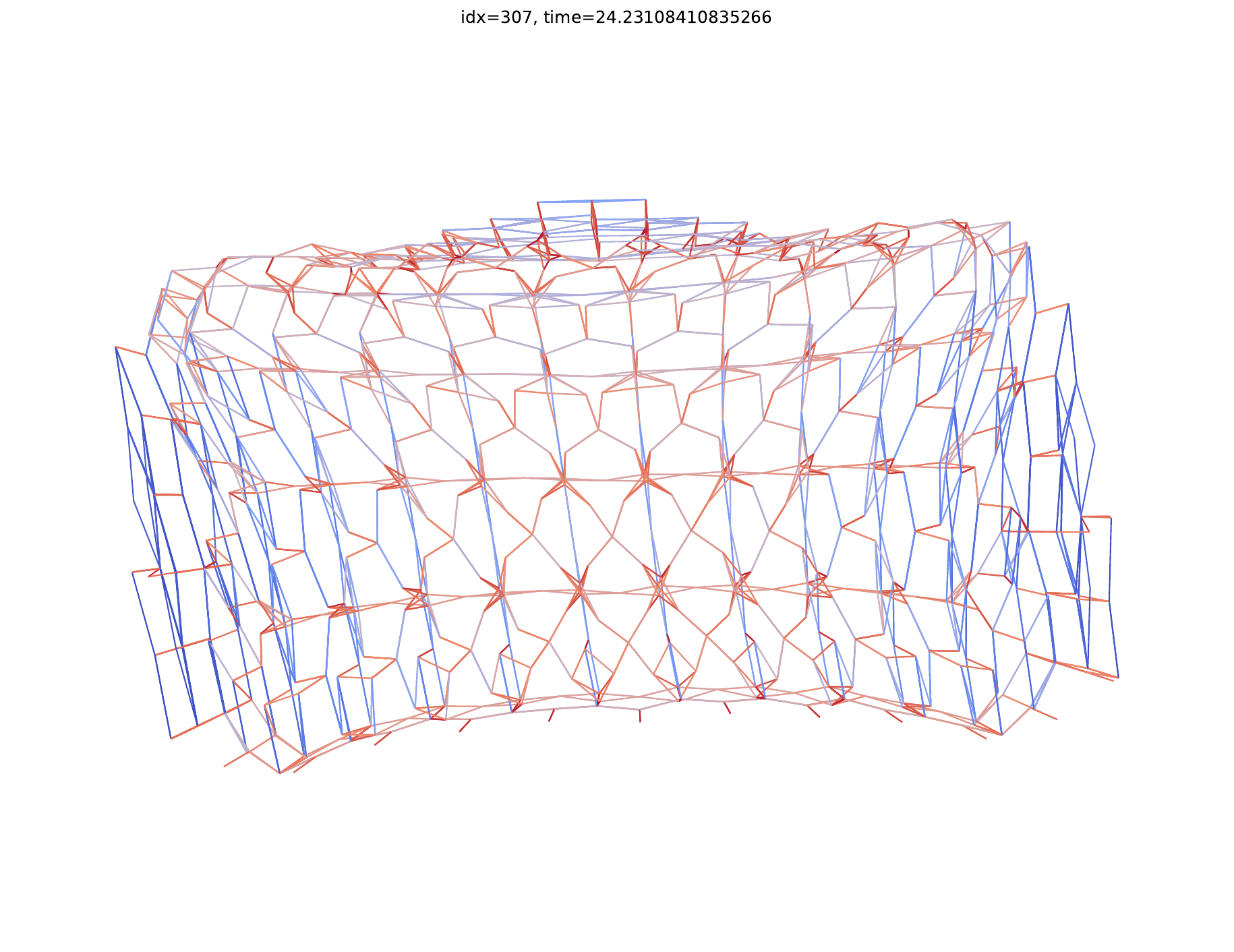} &
\imgcell{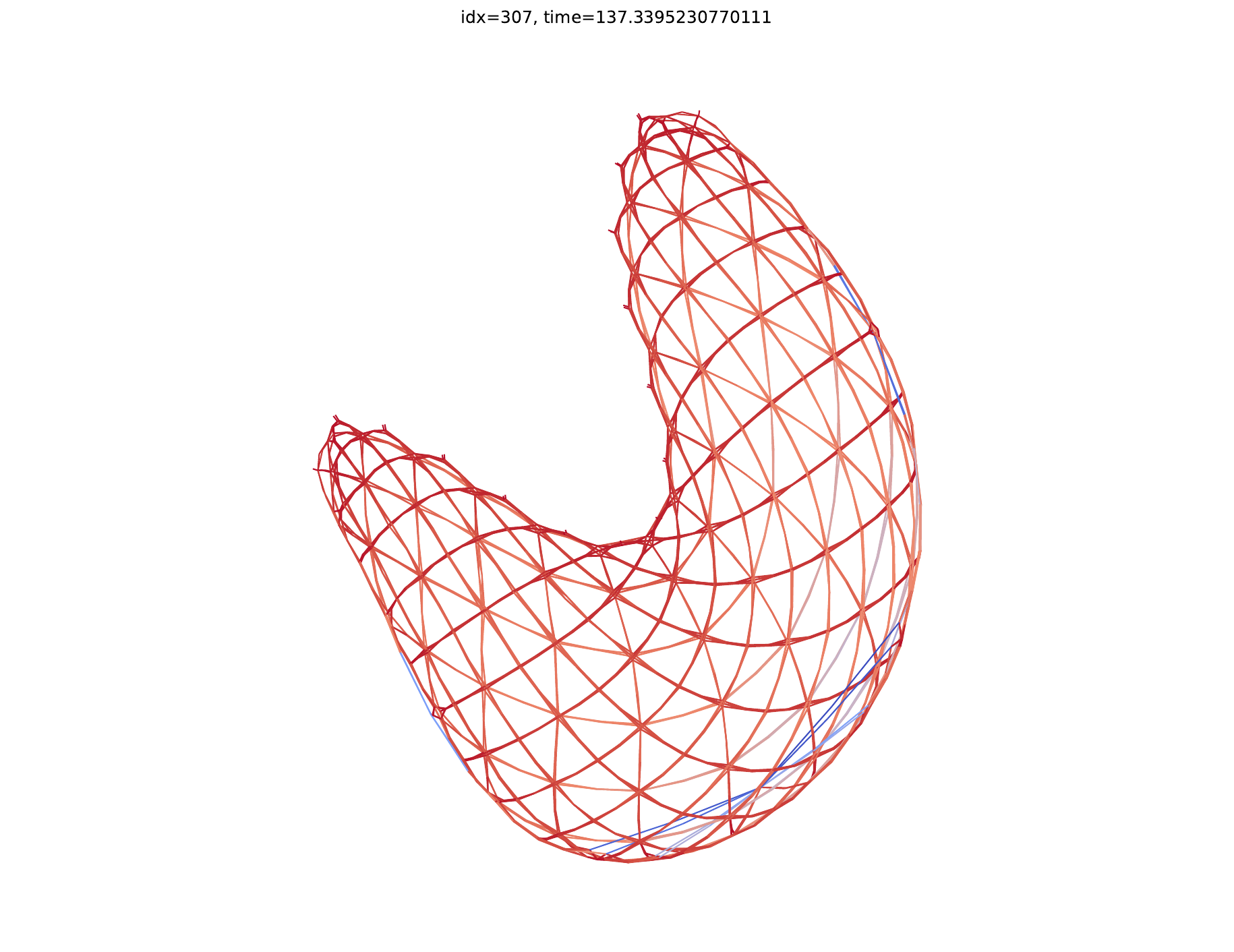} &
\imgcell{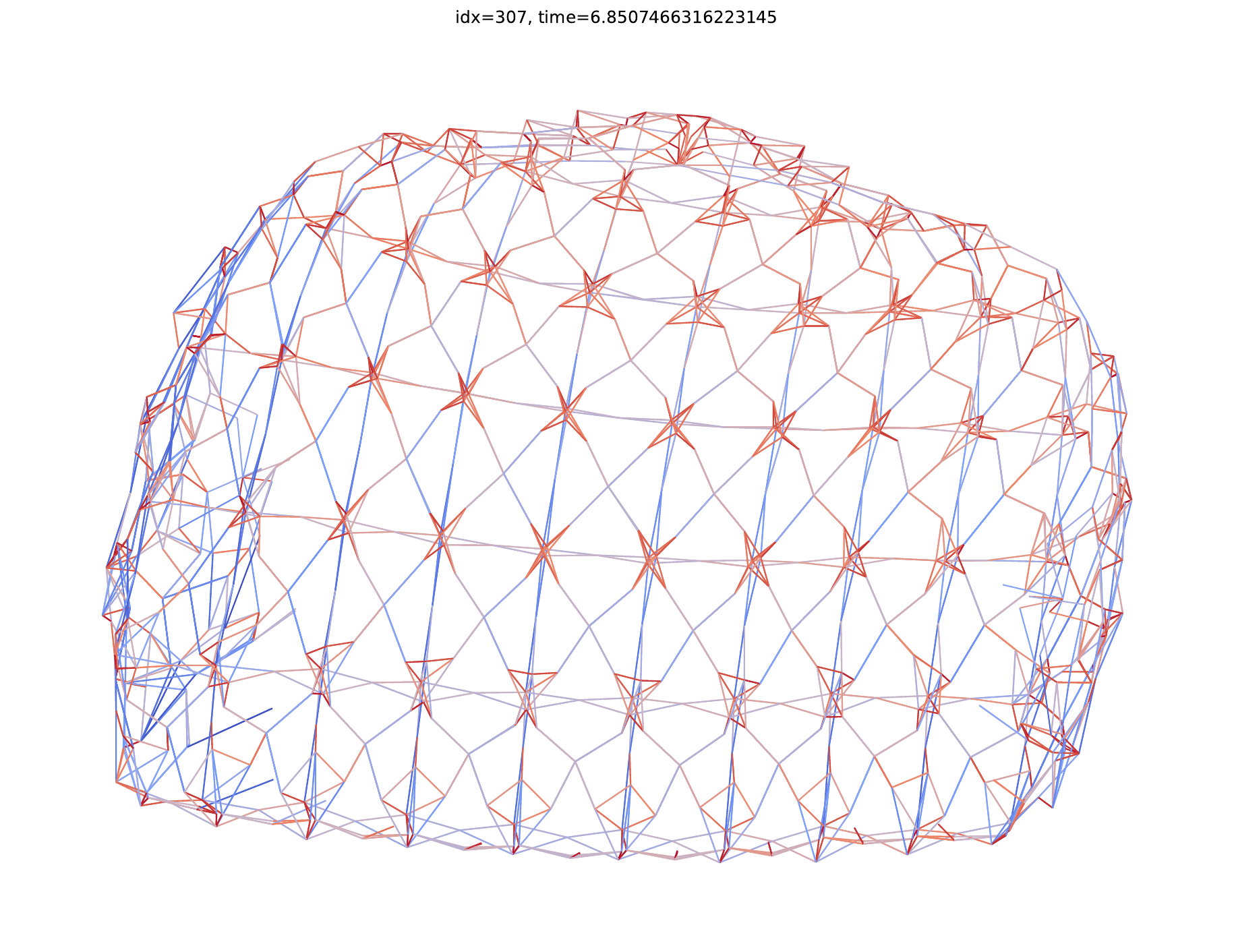} &
\imgcell{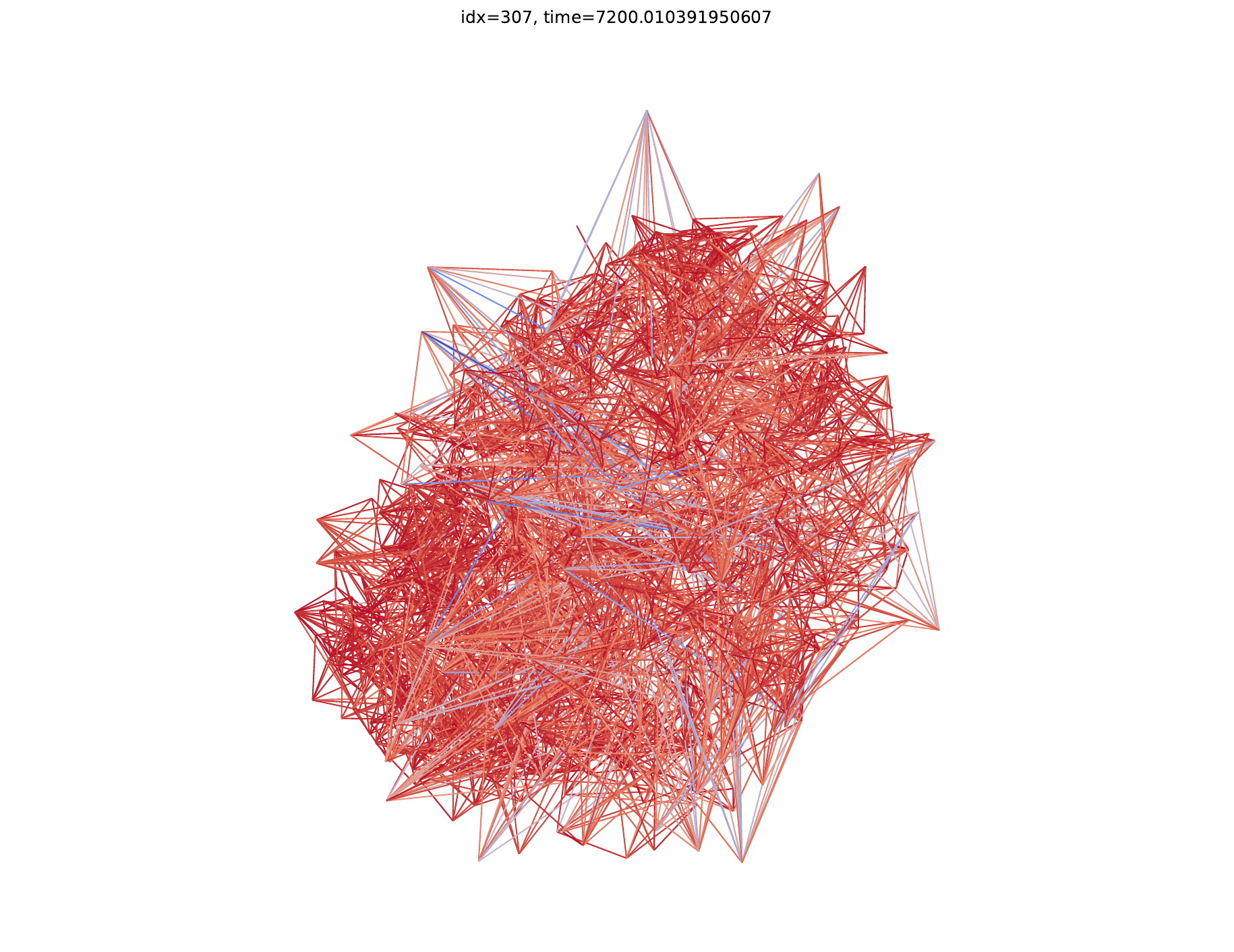} &
\imgcell{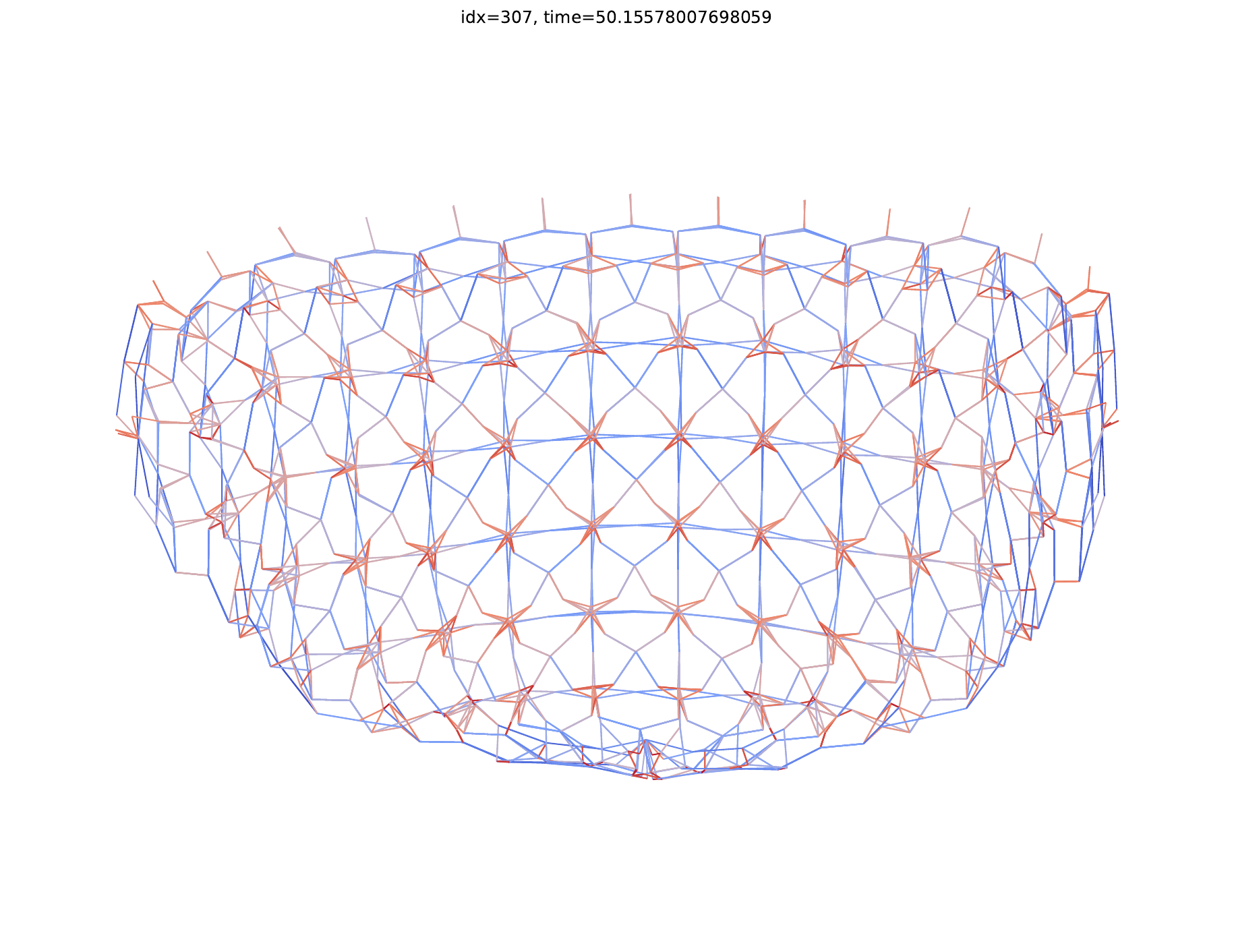} &
\imgcell{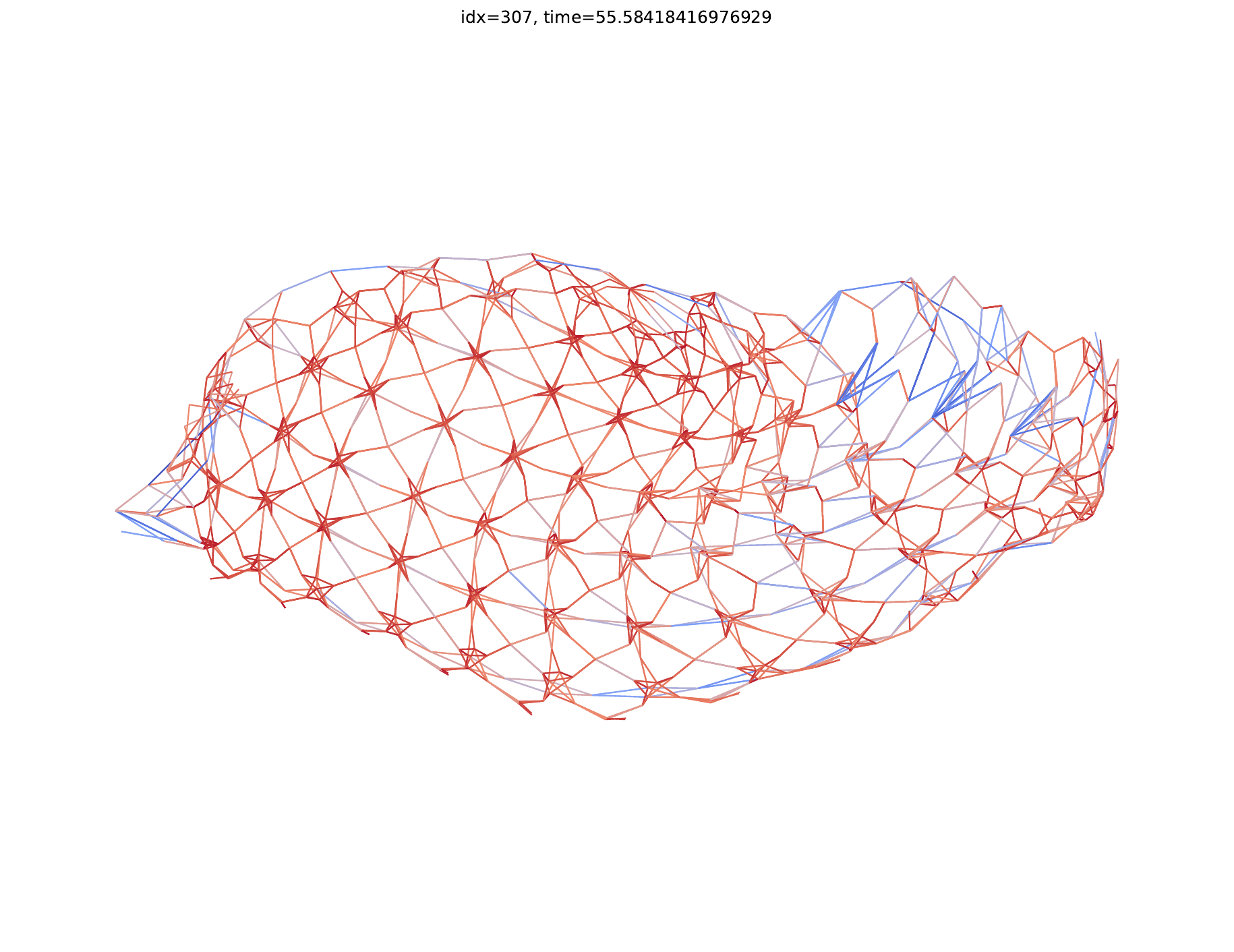} &
\imgcell{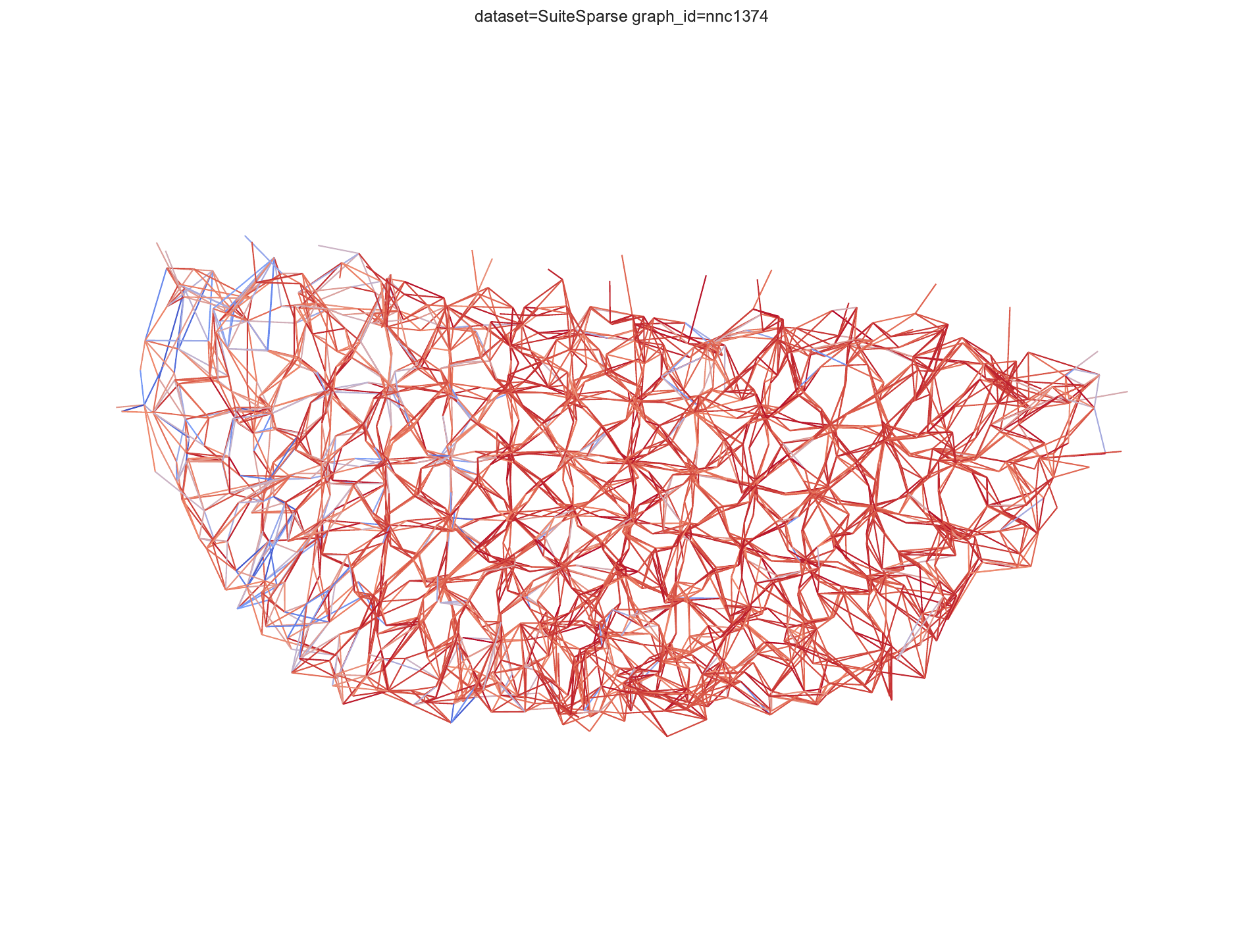} &
\imgcell{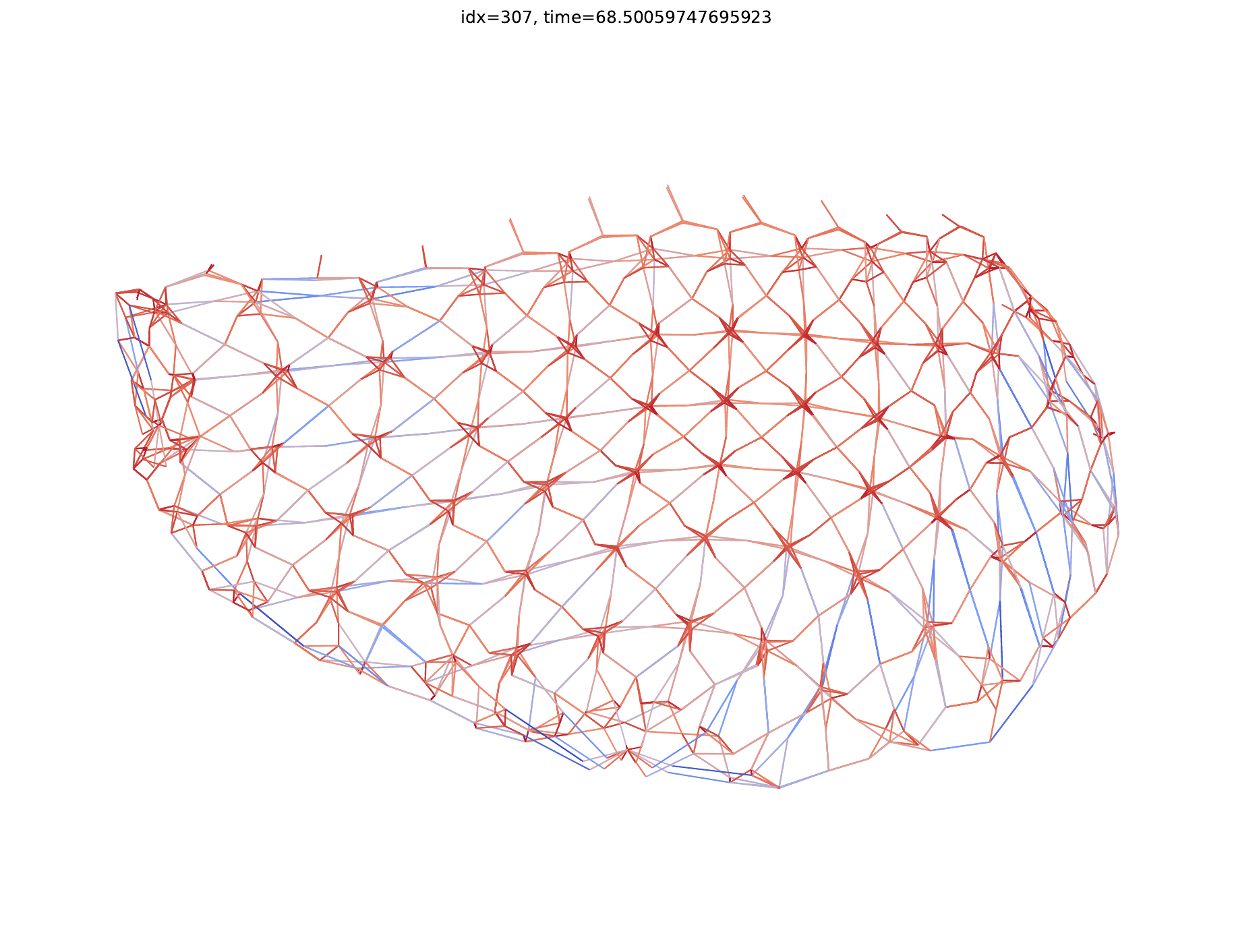} &
\imgcell{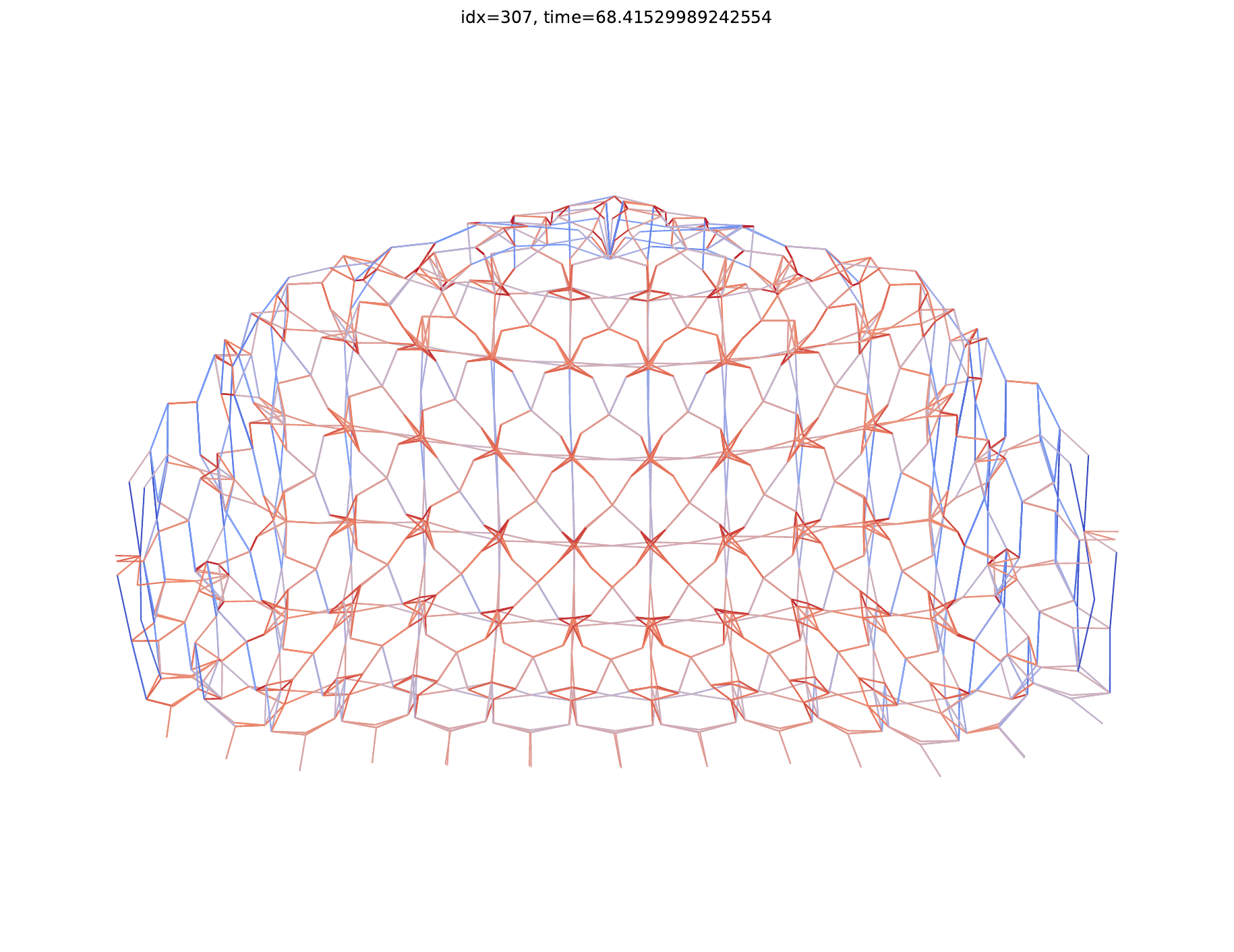} &
\imgcell{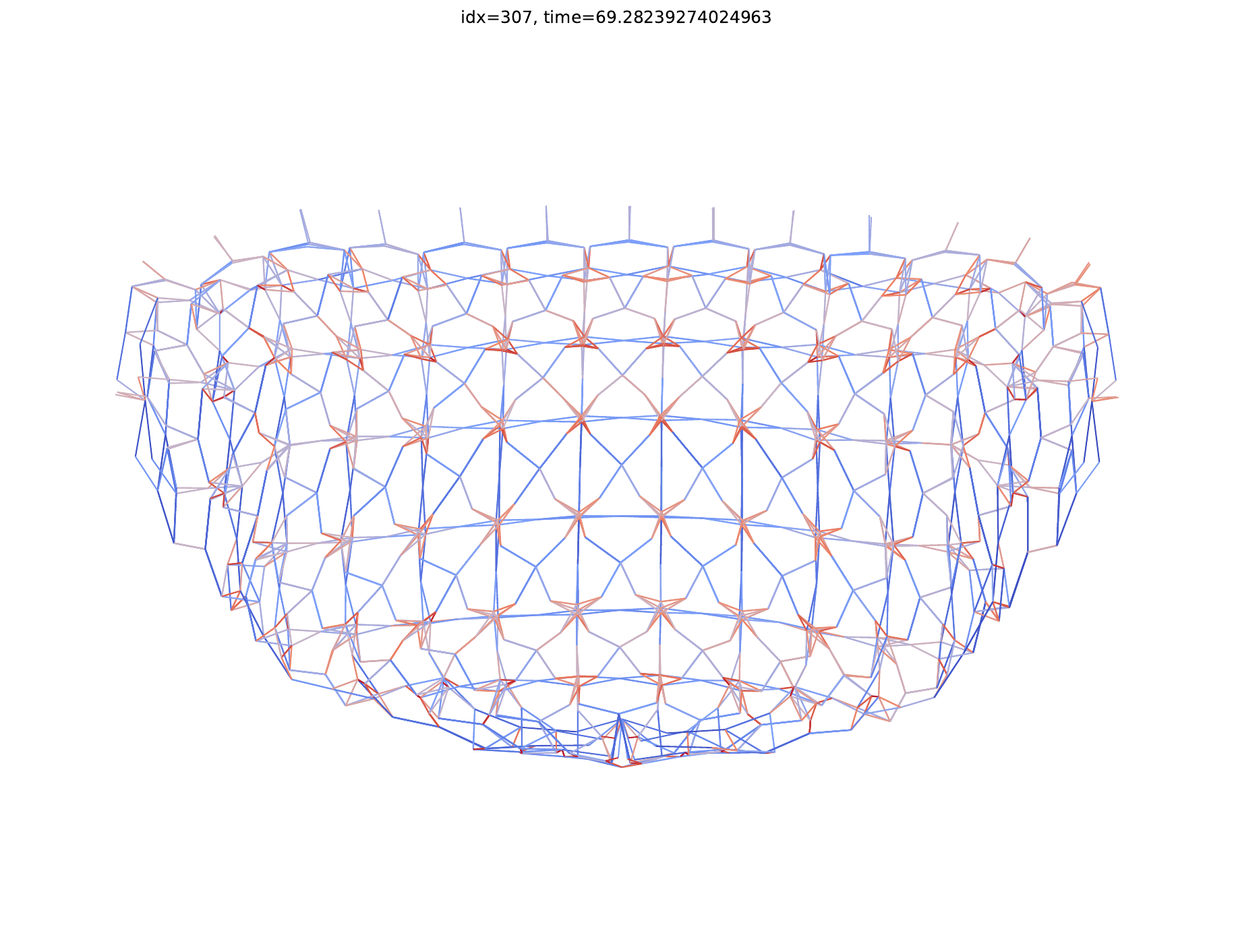} &
\imgcell{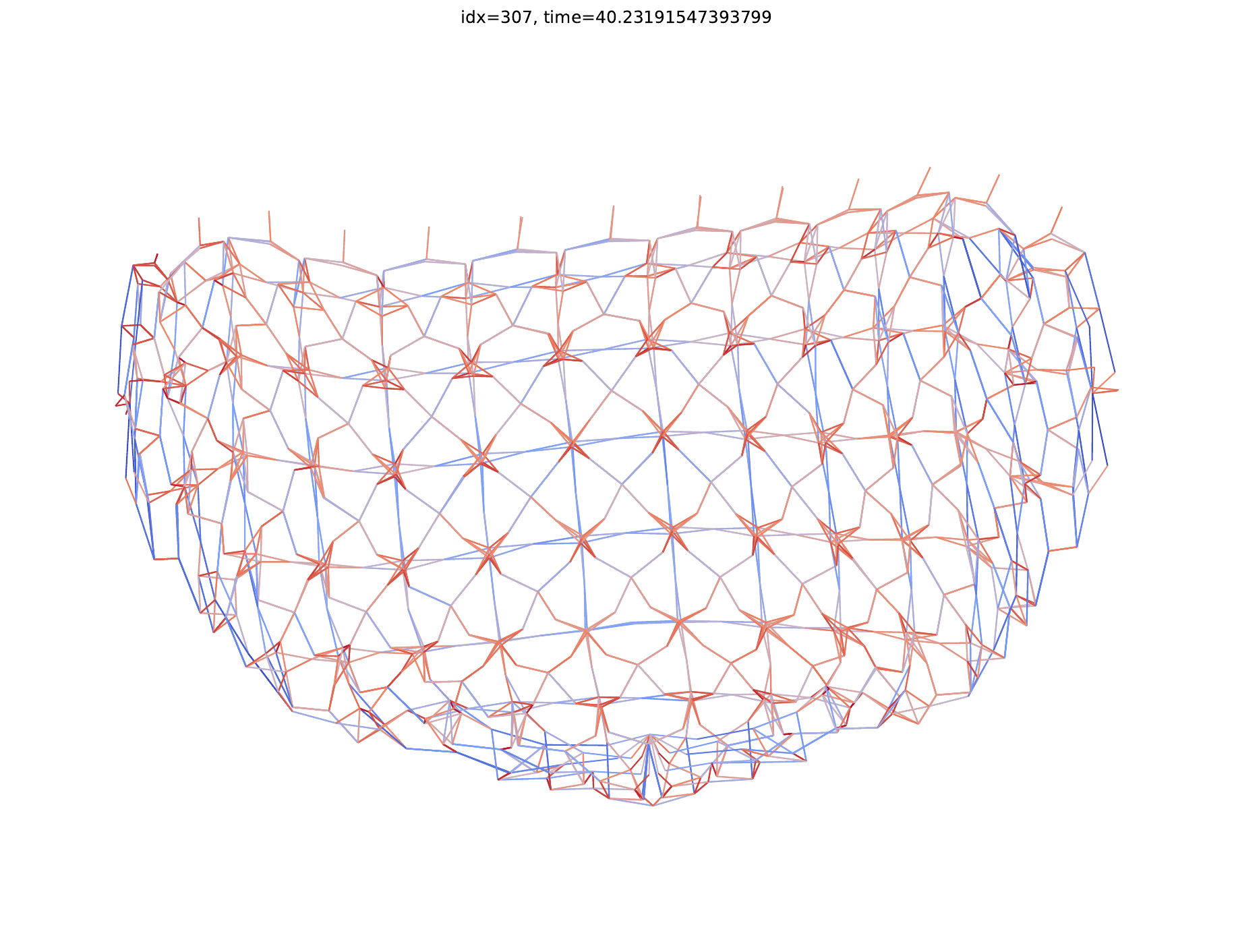} \\

&
t = 0.92s &
t = 24.23s &
t = 137.34s &
t = 6.85s &
t = 7200.00s &
t = 7.06s &
t = 6.58s &
t = 7.12s &
t = 6.80s &
t = 6.82s &
t = 6.98s &
t = 6.70s \\

\makecell{\bfseries ex4\\N = 1601\\M = 15349} &
\imgcell{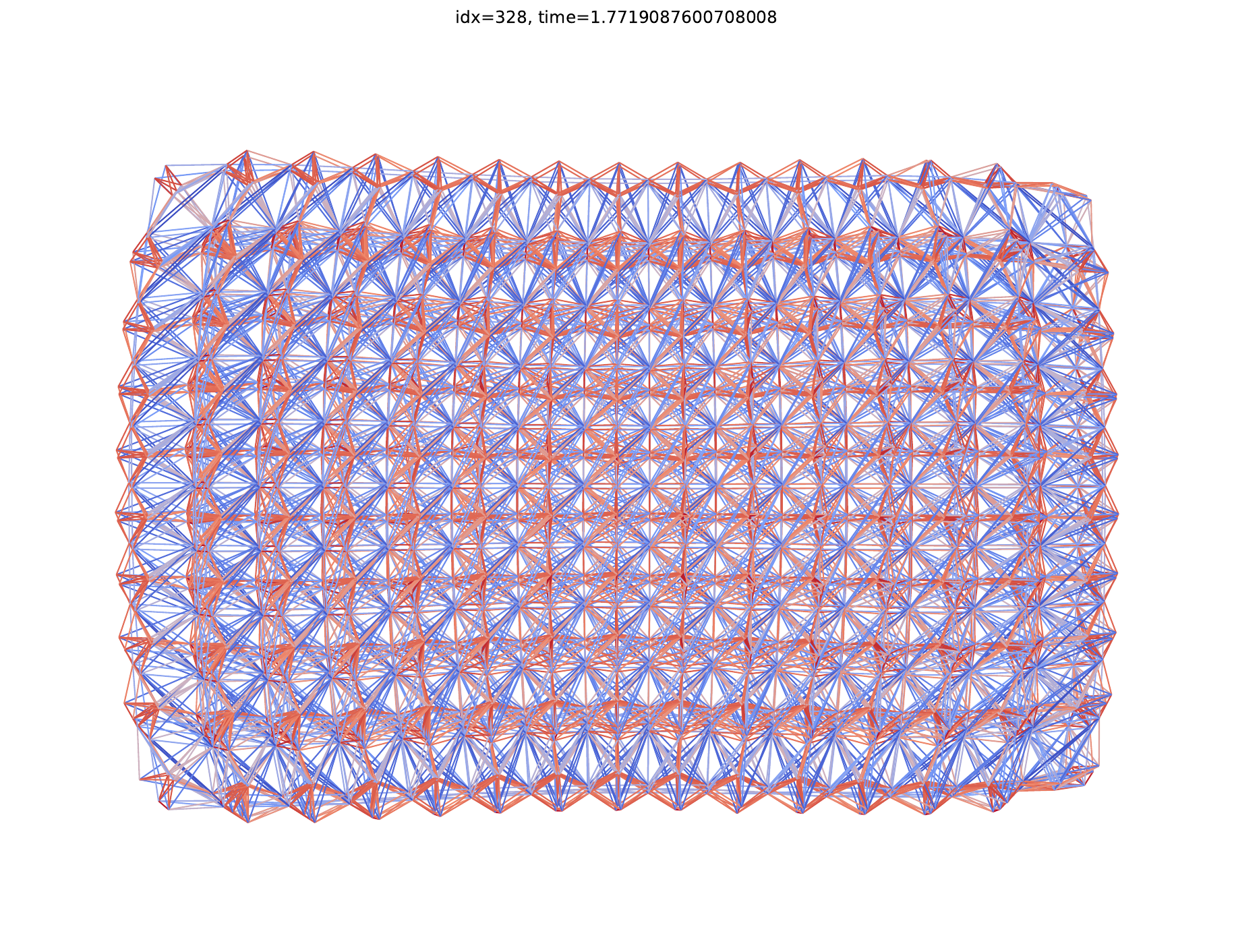} &
\imgcell{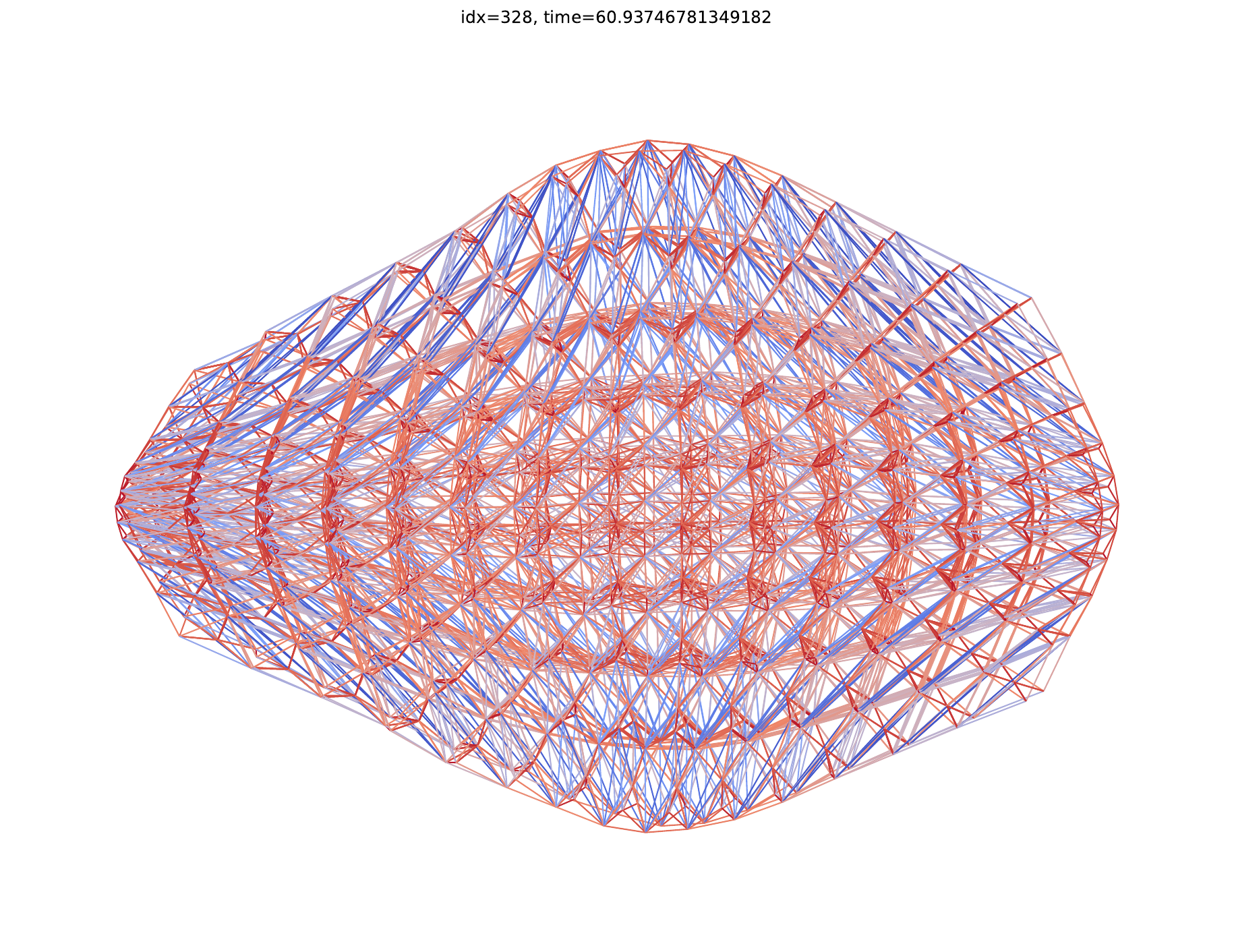} &
\imgcell{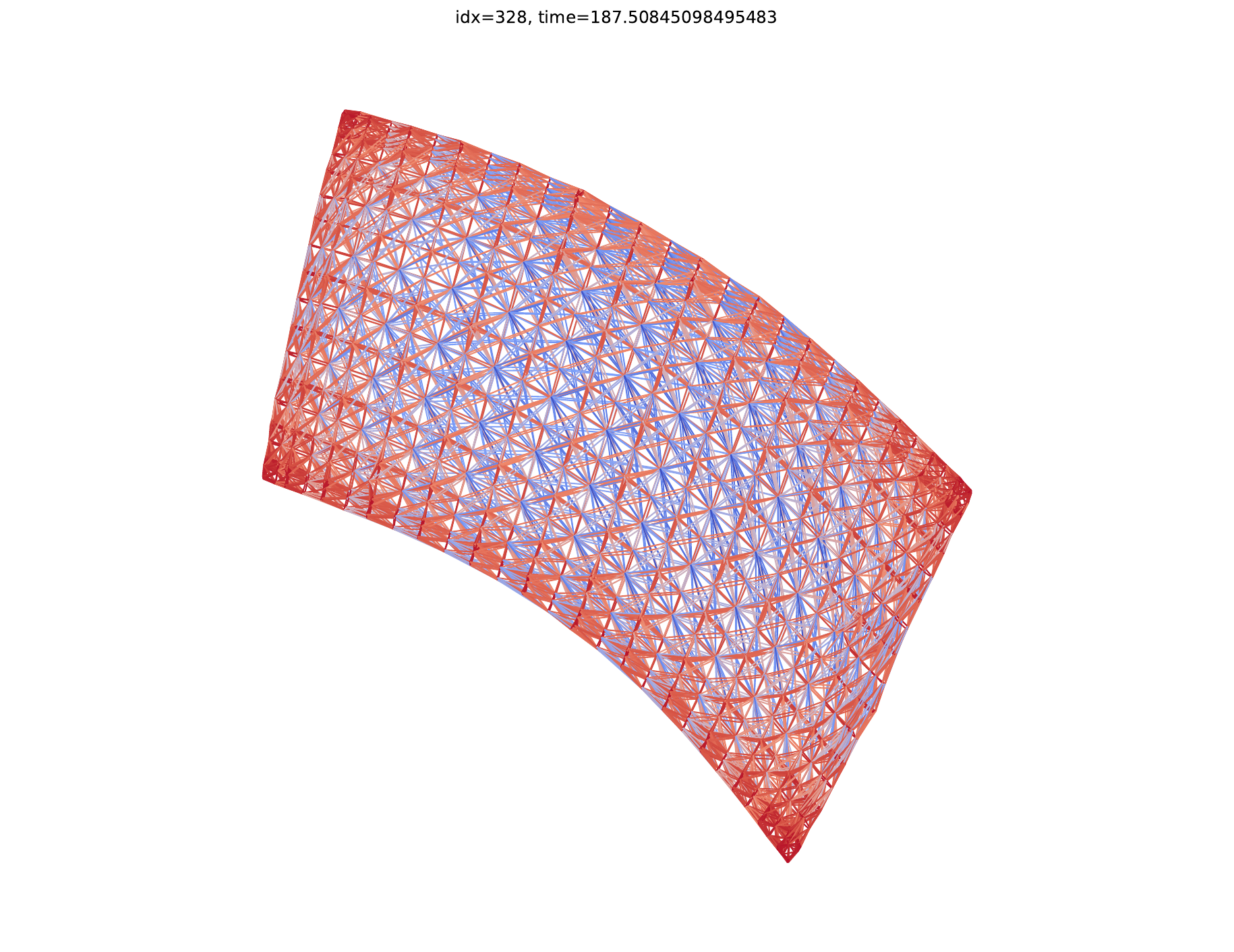} &
\imgcell{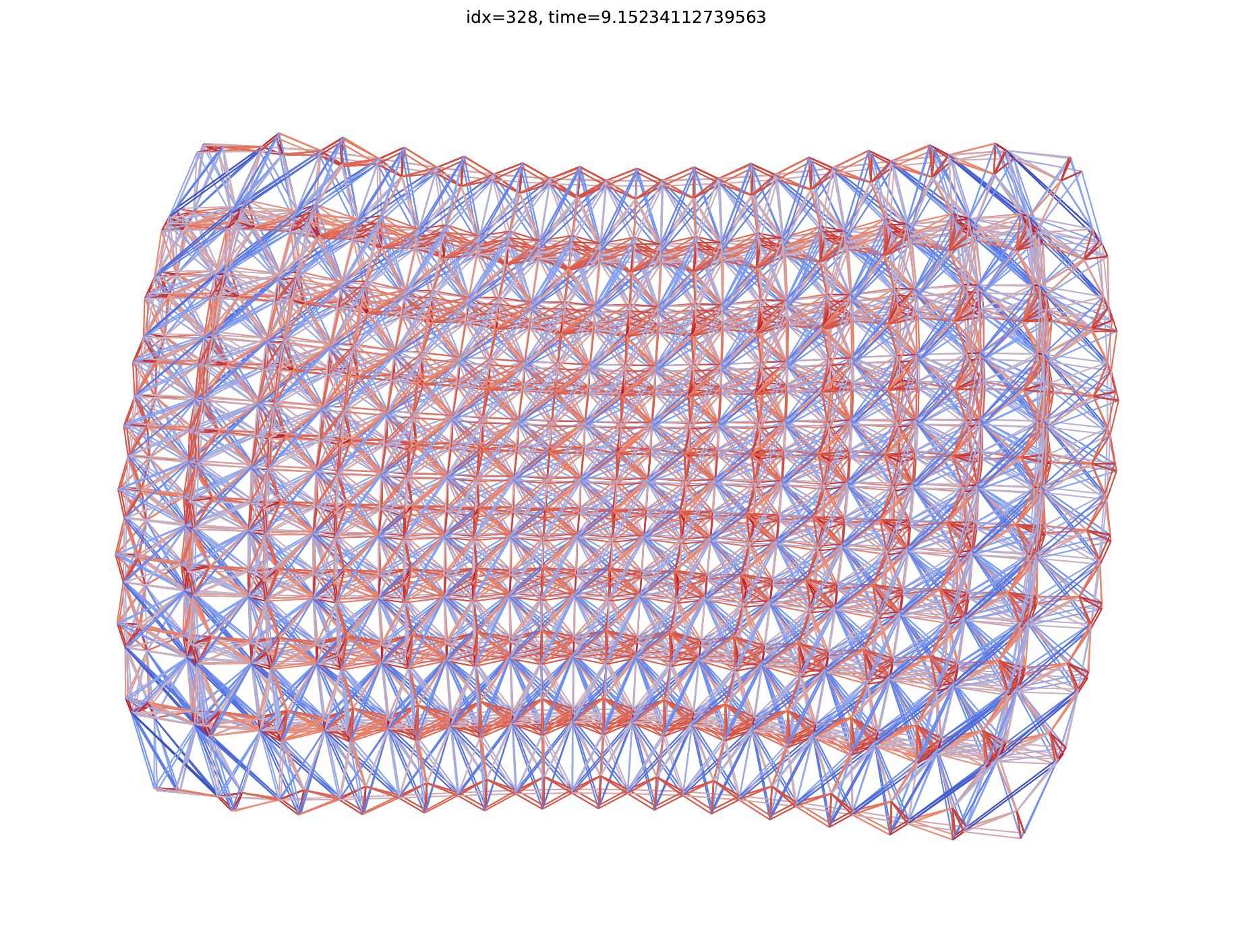} &
\imgcell{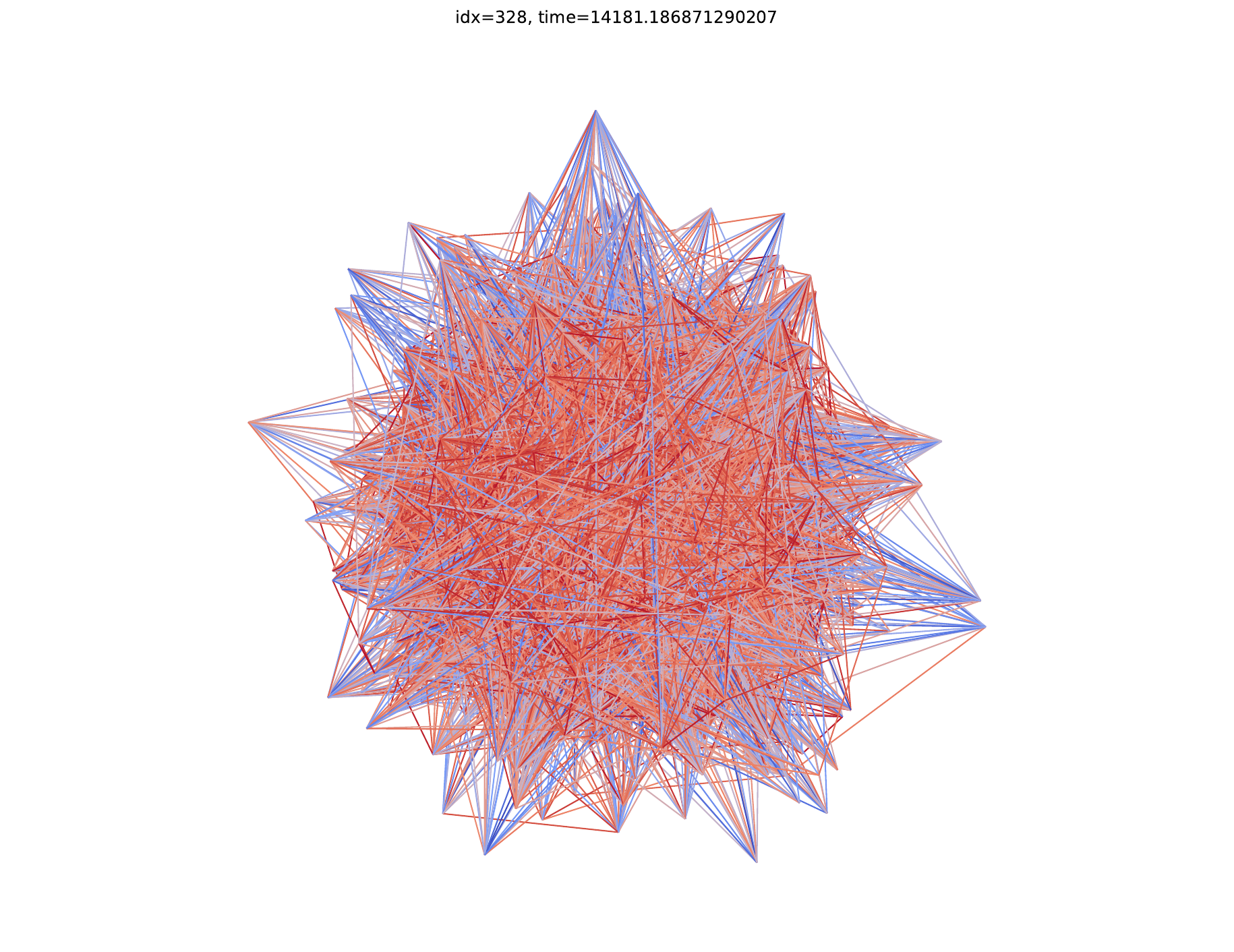} &
\imgcell{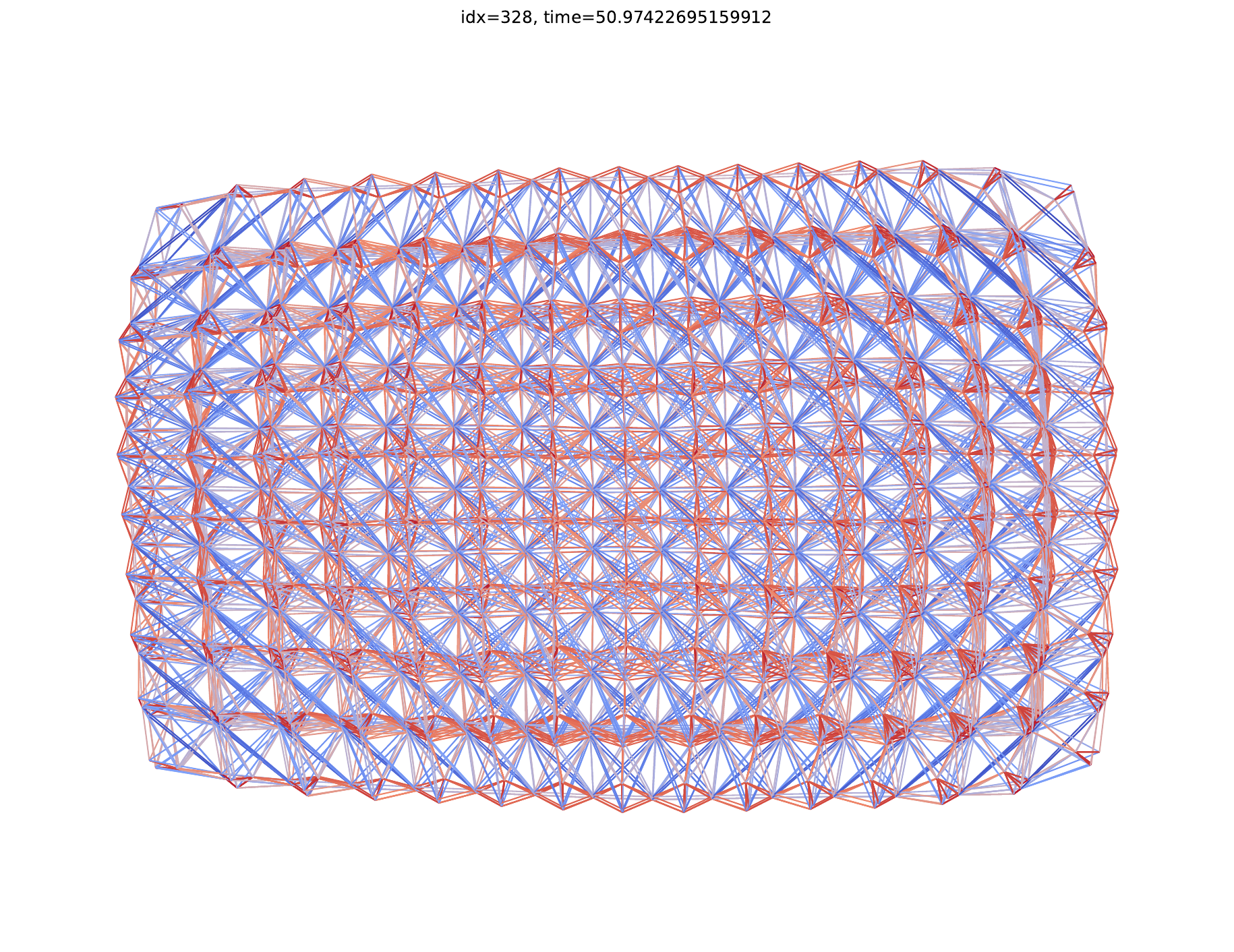} &
\imgcell{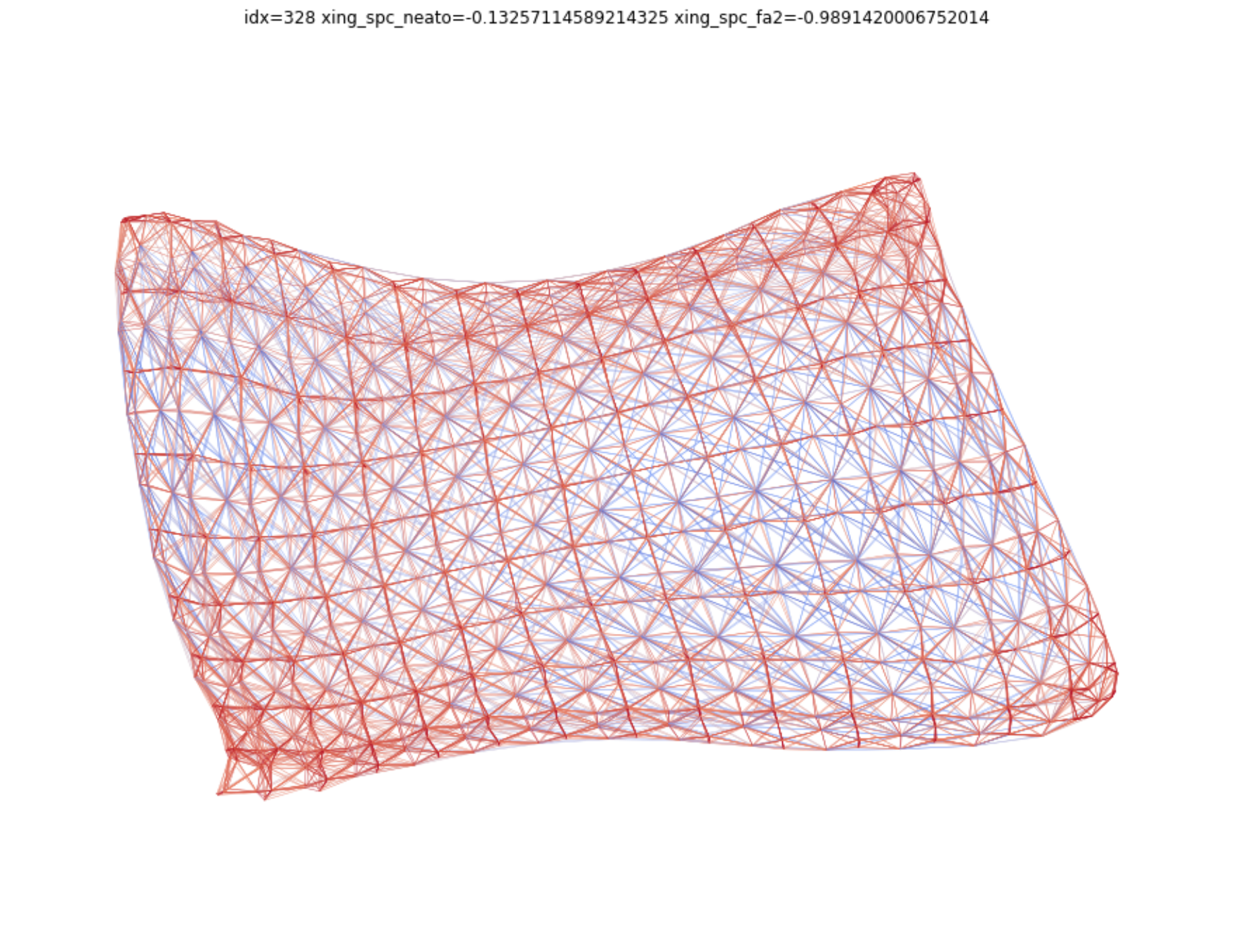} &
\imgcell{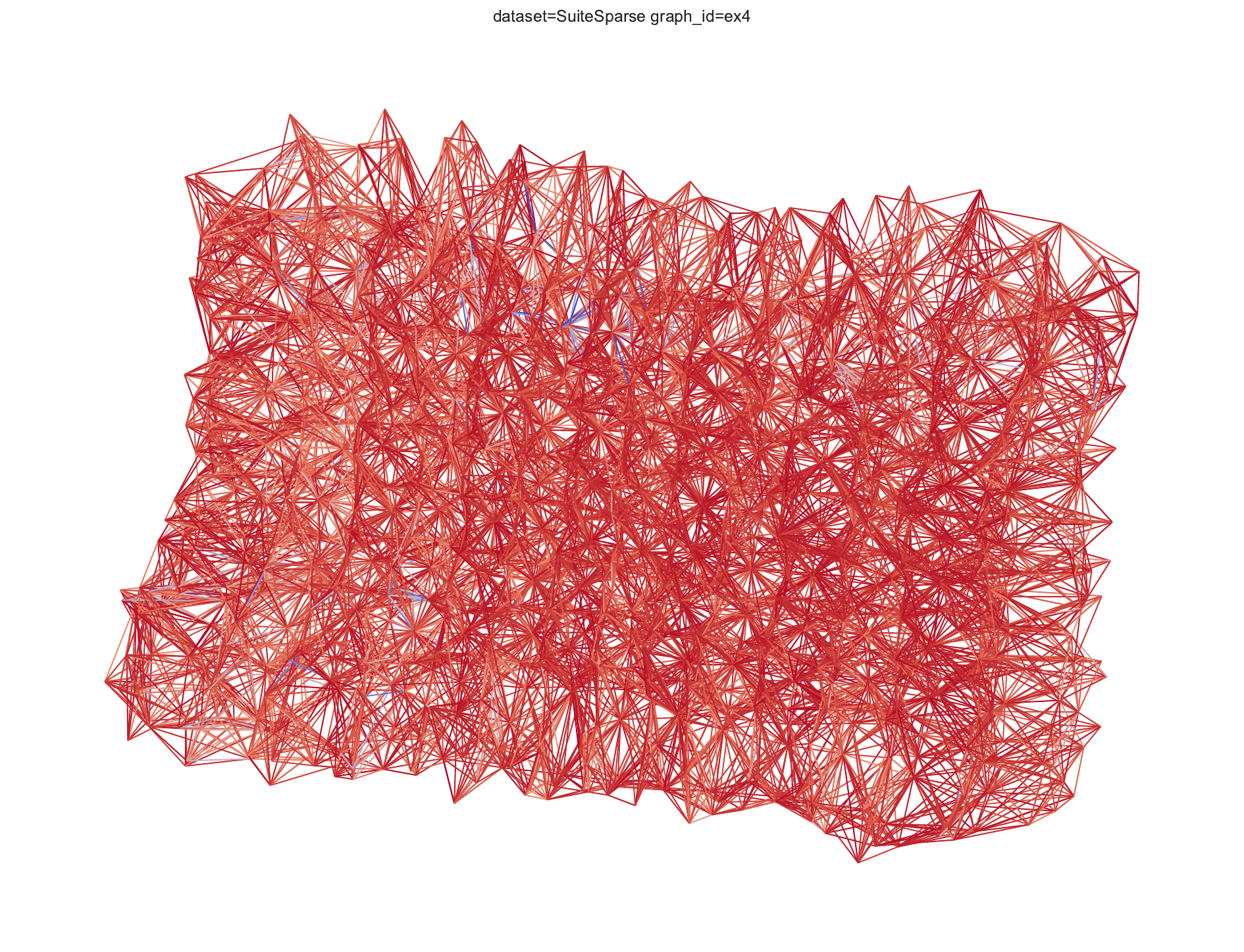} &
\imgcell{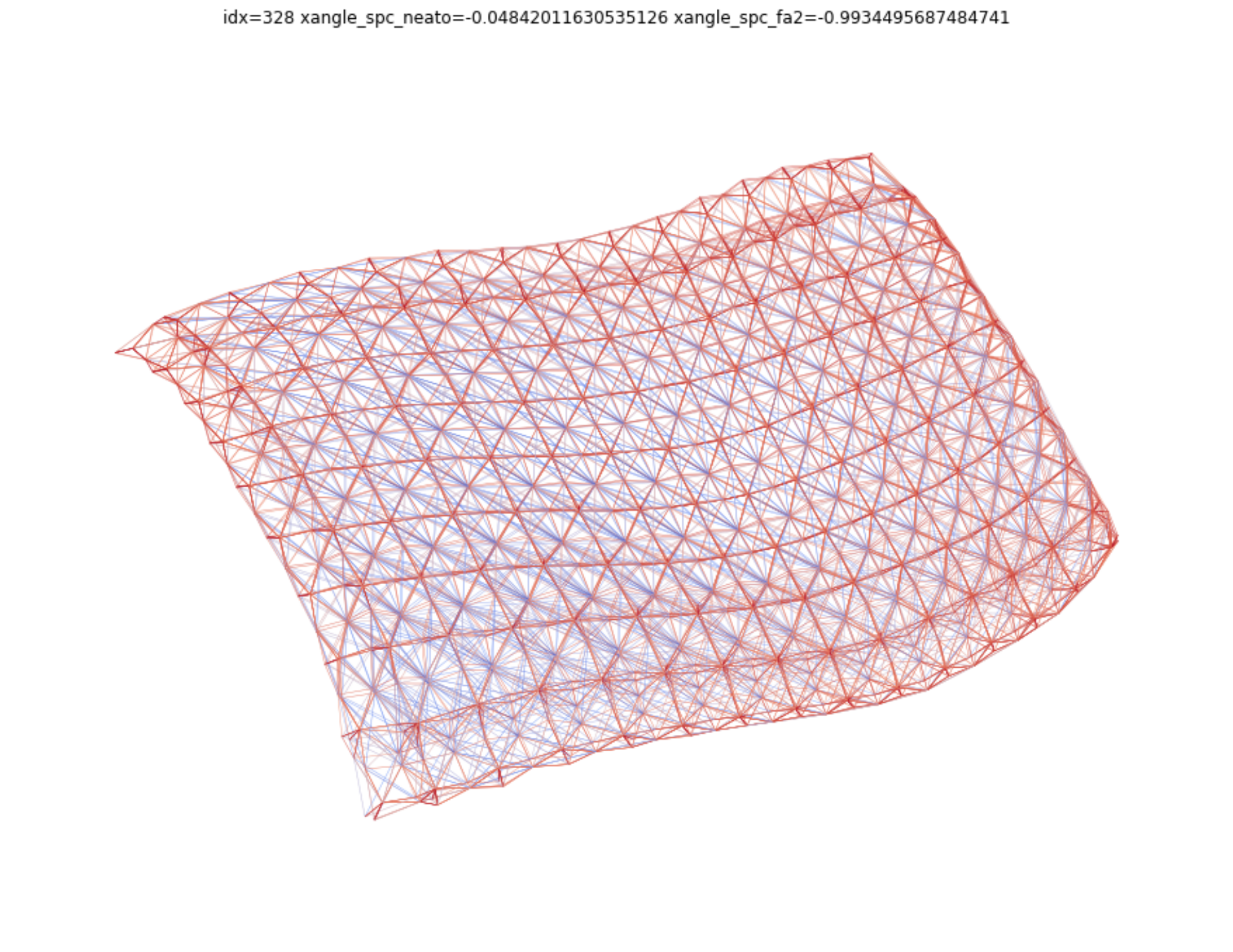} &
\imgcell{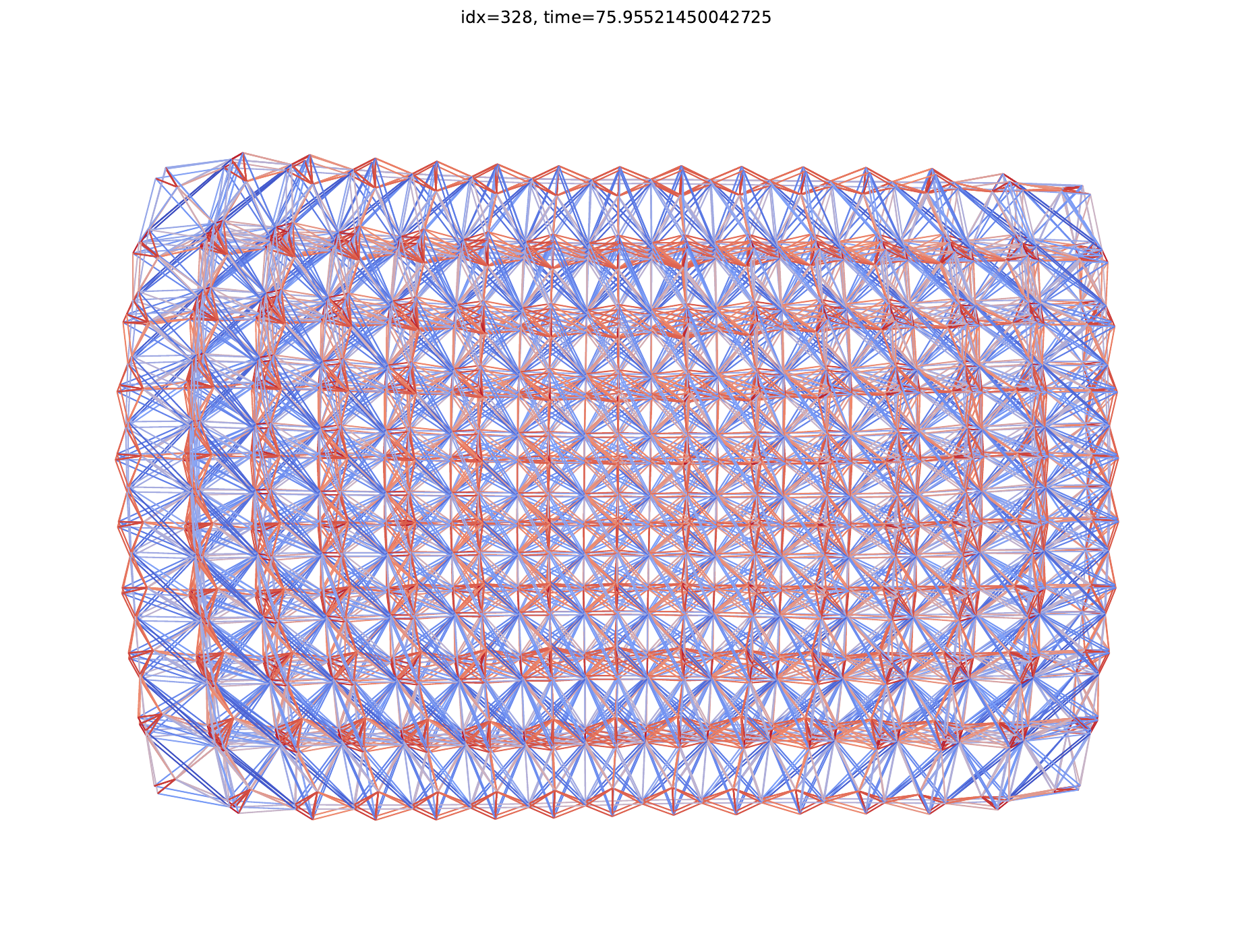} &
\imgcell{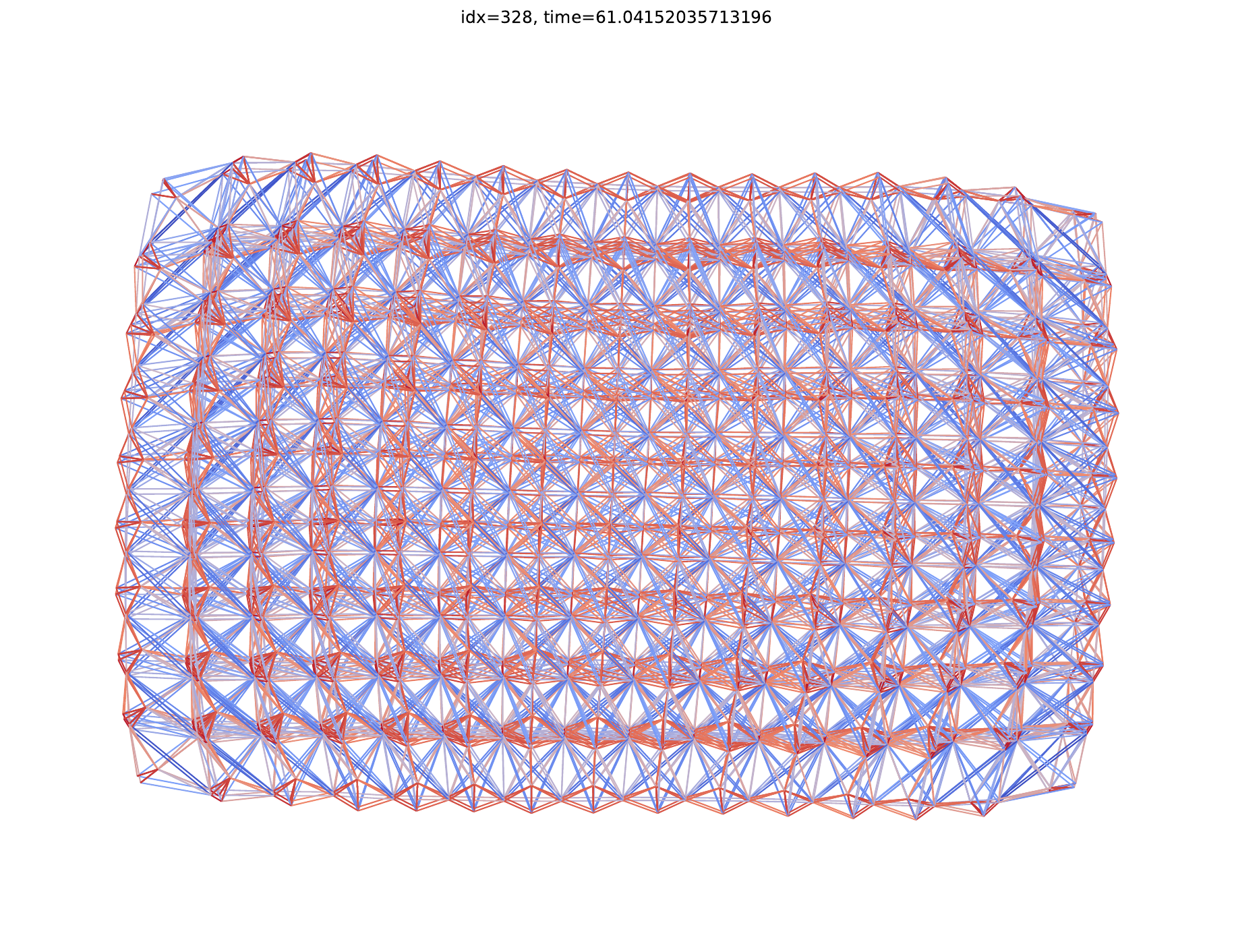} &
\imgcell{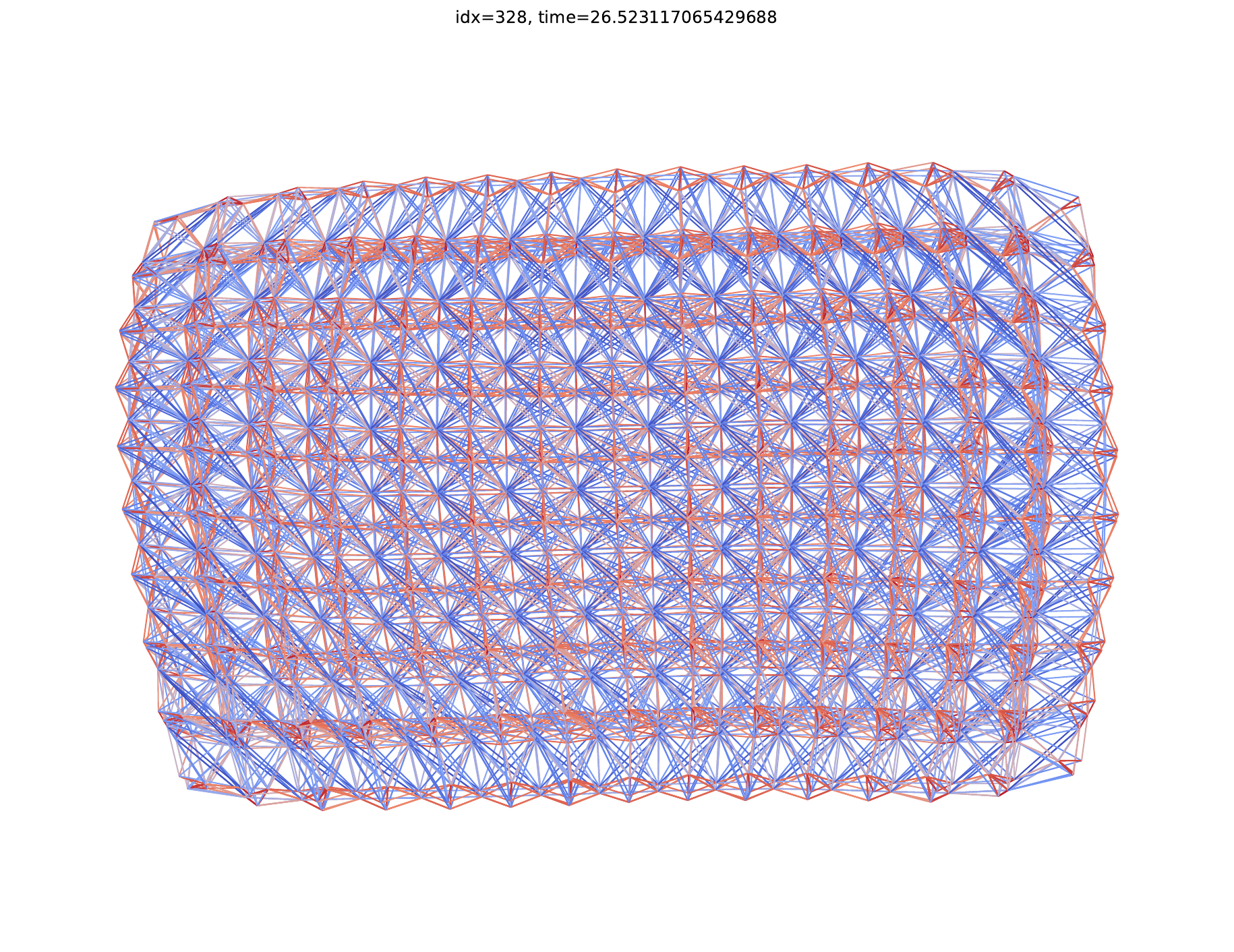} \\

&
t = 1.77s &
t = 60.94s &
t = 187.51s &
t = 9.15s &
t = 7200.00s &
t = 10.01s &
t = 9.41s &
t = 10.24s &
t = 8.75s &
t = 8.96s &
t = 11.04s &
t = 9.44s \\

\makecell{\bfseries utm1700b\\N = 1700\\M = 14626} &
\imgcell{figures/large_graphs/336_sgd2.pdf} &
\imgcell{figures/large_graphs/336_pmds.pdf} &
\imgcell{figures/large_graphs/336_fa2.pdf} &
\imgcell{figures/large_graphs/336_deepgd.pdf} &
\imgcell{figures/large_graphs/336_gd2_stress_xing.pdf} &
\imgcell{figures/large_graphs/336_smartgd_stress.pdf} &
\imgcell{figures/large_graphs/336_smartgd_xing.pdf} &
\imgcell{figures/large_graphs/336_smartgd_xing_nsc.pdf} &
\imgcell{figures/large_graphs/336_smartgd_xangle.pdf} &
\imgcell{figures/large_graphs/336_smartgd_stress_xing.pdf} &
\imgcell{figures/large_graphs/336_smartgd_stress_xangle.pdf} &
\imgcell{figures/large_graphs/336_smartgd_combined.pdf} \\

&
t = 2.02s &
t = 243.70s &
t = 227.30s &
t = 10.38s &
t = 7200.00s &
t = 10.49s &
t = 8.85s &
t = 10.77s &
t = 9.96s &
t = 9.95s &
t = 10.68s &
t = 11.99s \\

\makecell{\bfseries g7jac010sc\\N = 2880\\M = 16160} &
\imgcell{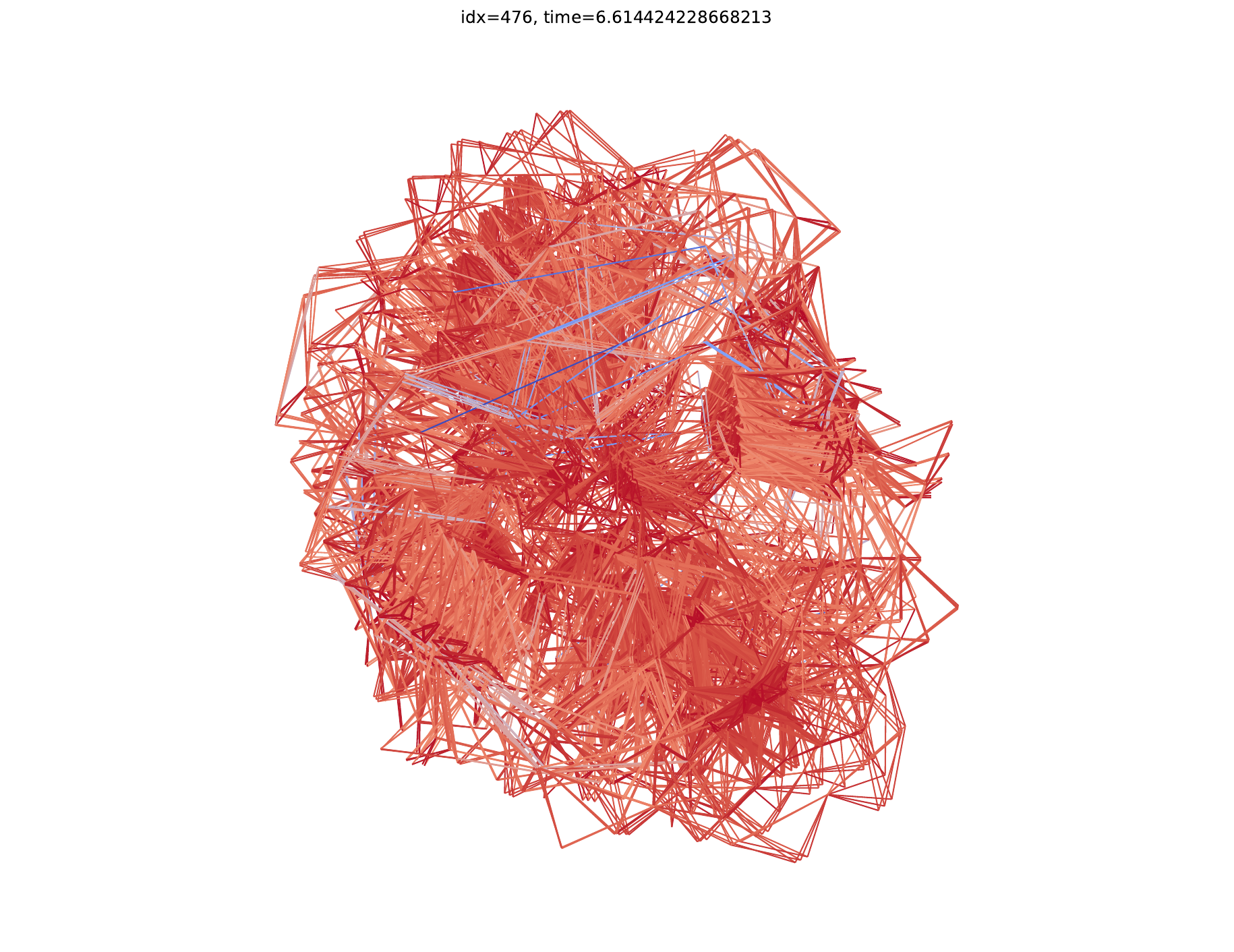} &
\imgcell{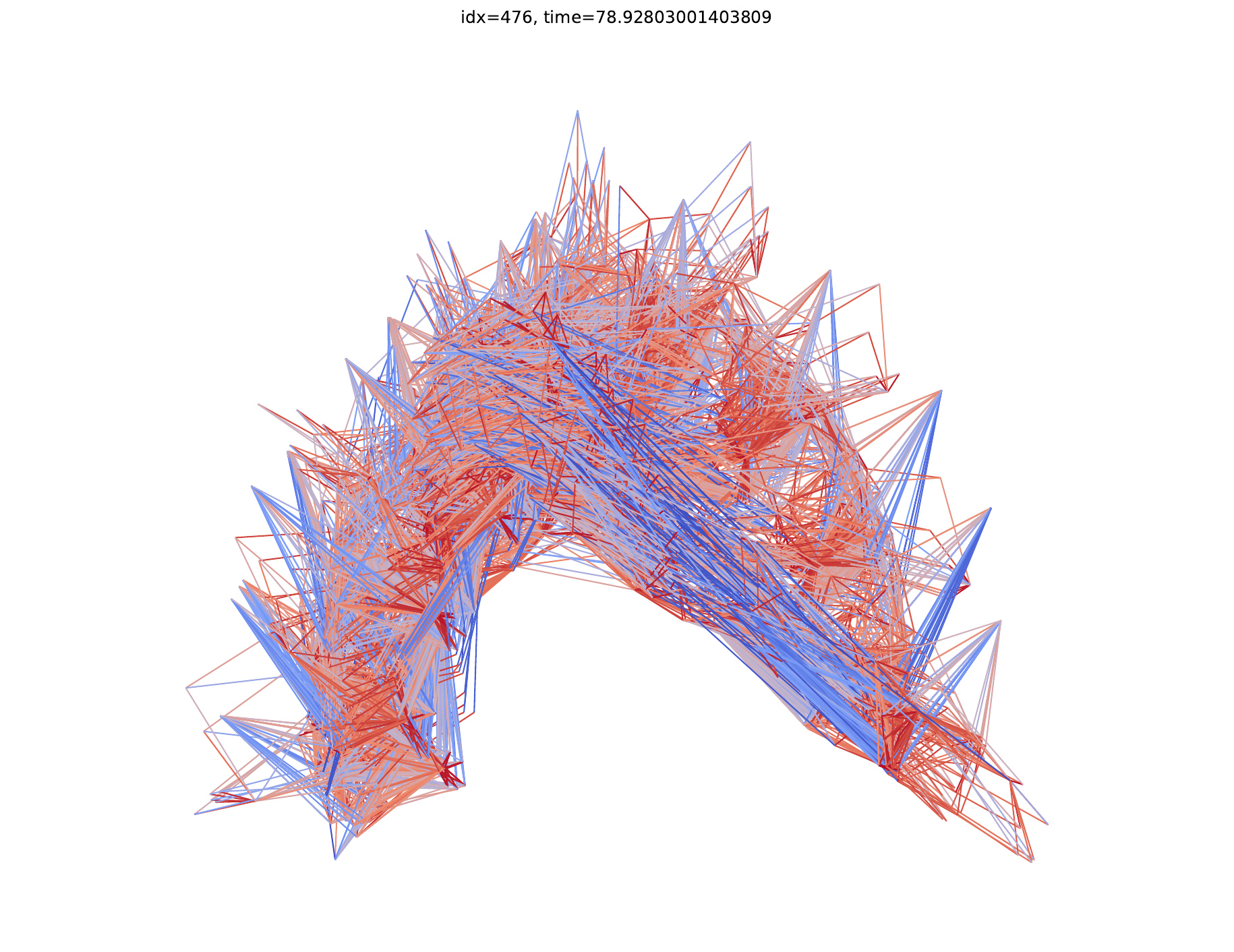} &
\imgcell{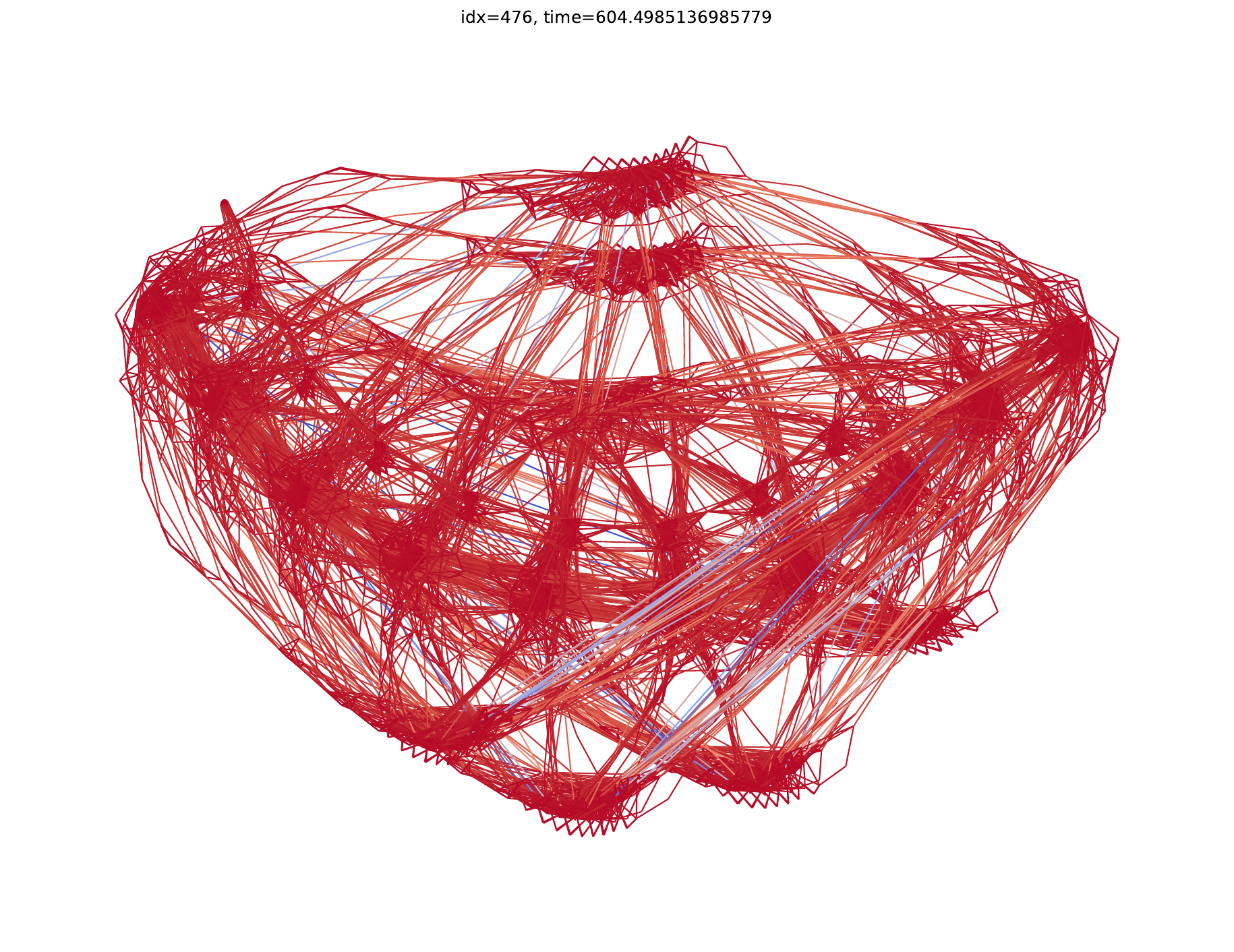} &
\imgcell{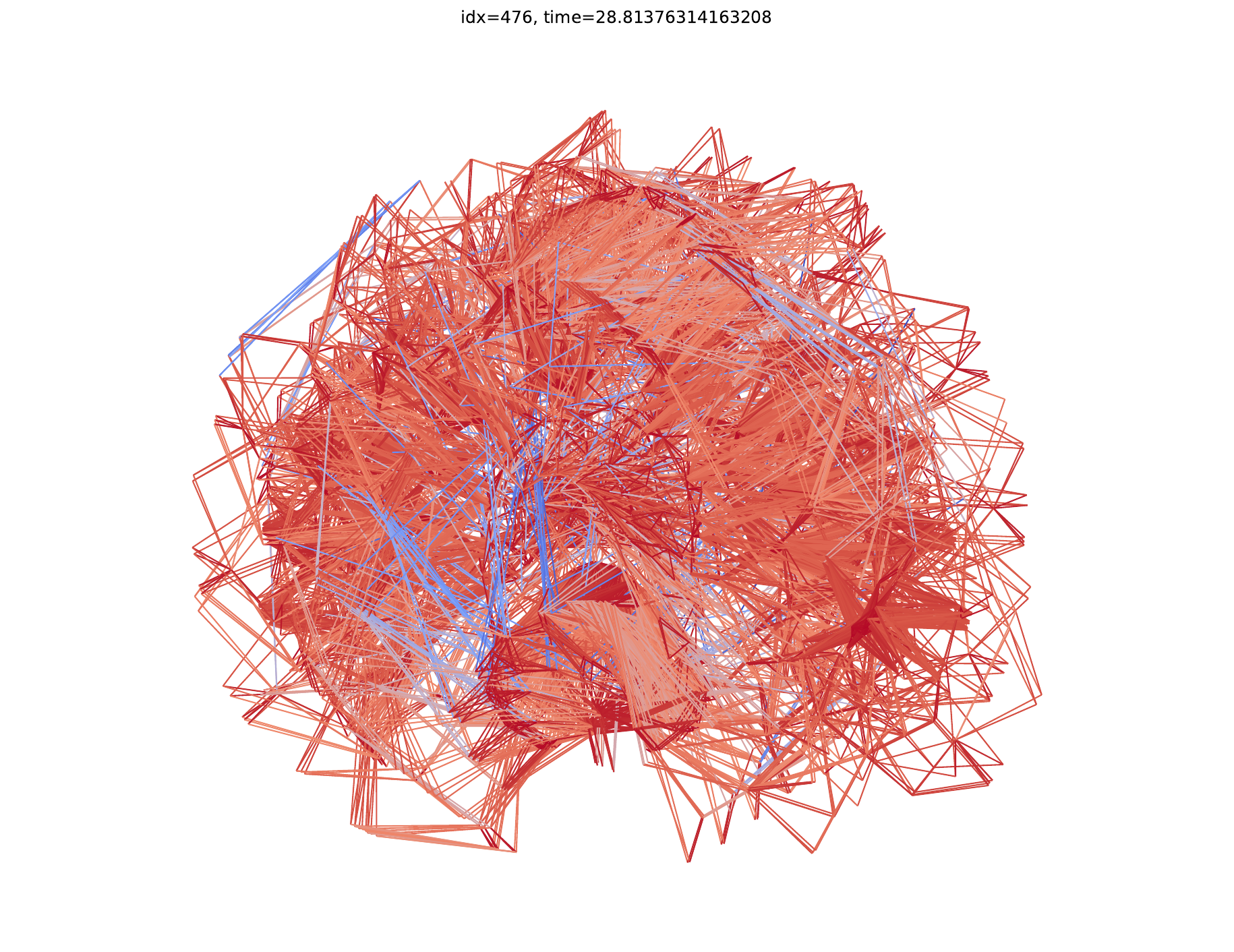} &
\imgcell{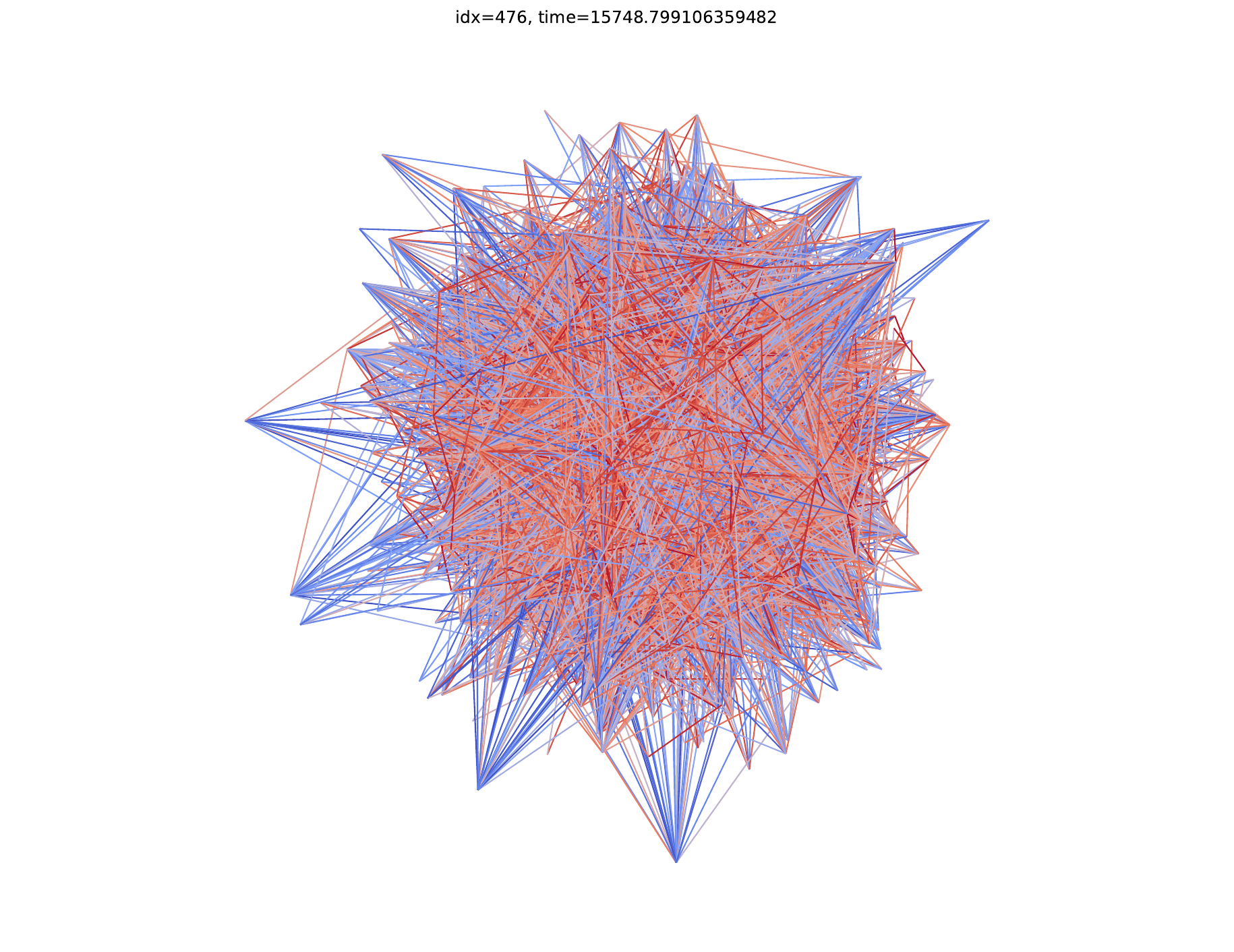} &
\imgcell{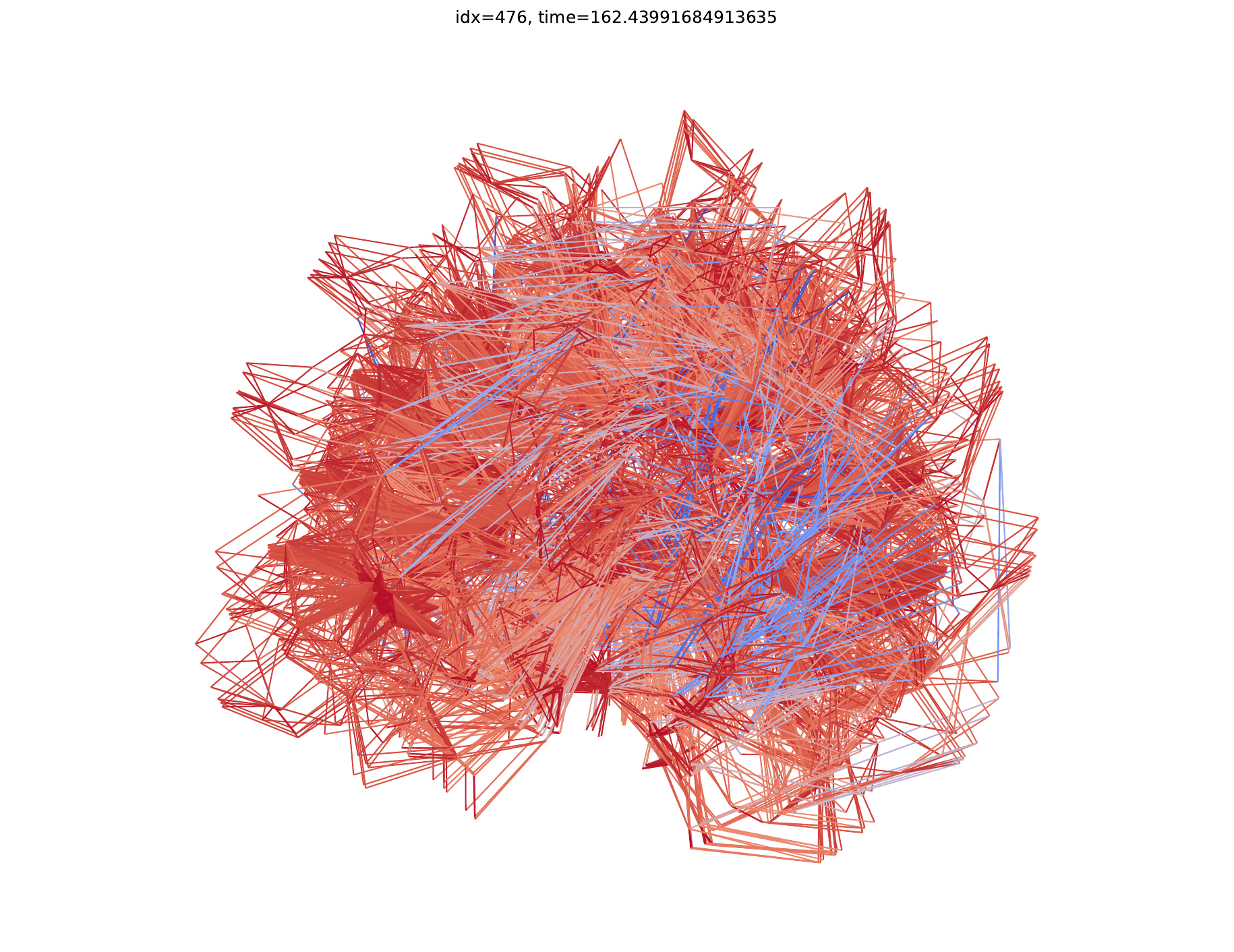} &
\imgcell{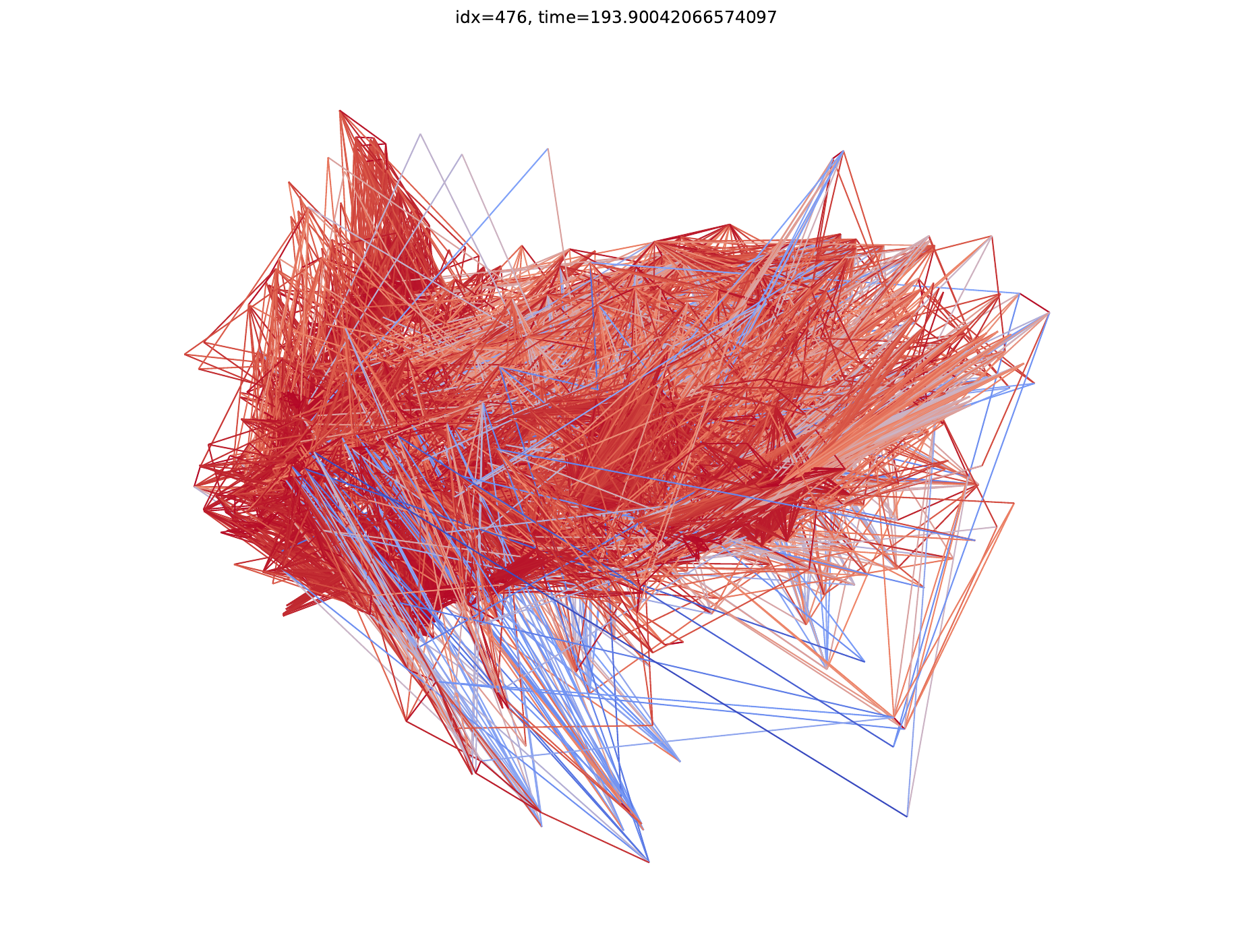} &
\imgcell{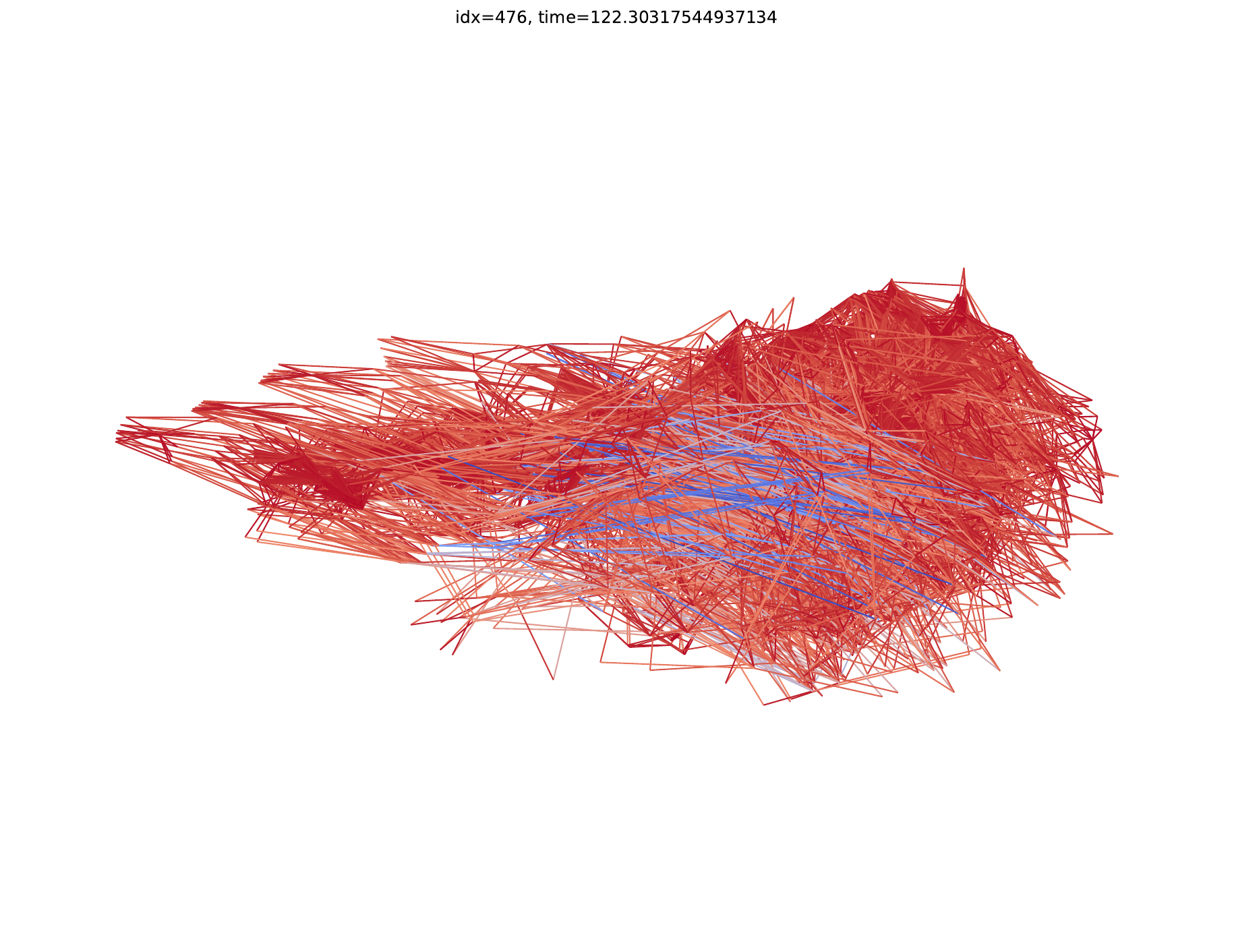} &
\imgcell{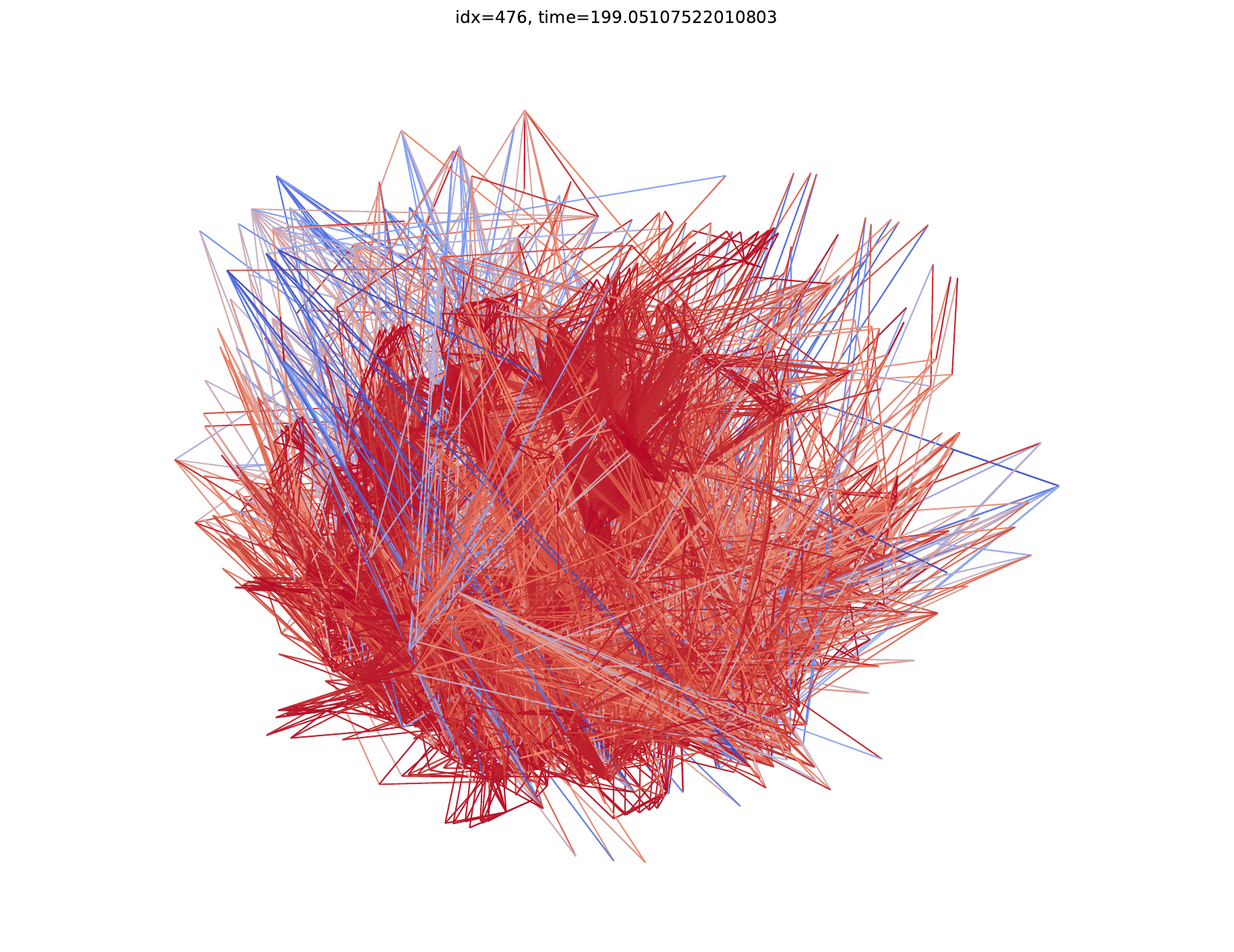} &
\imgcell{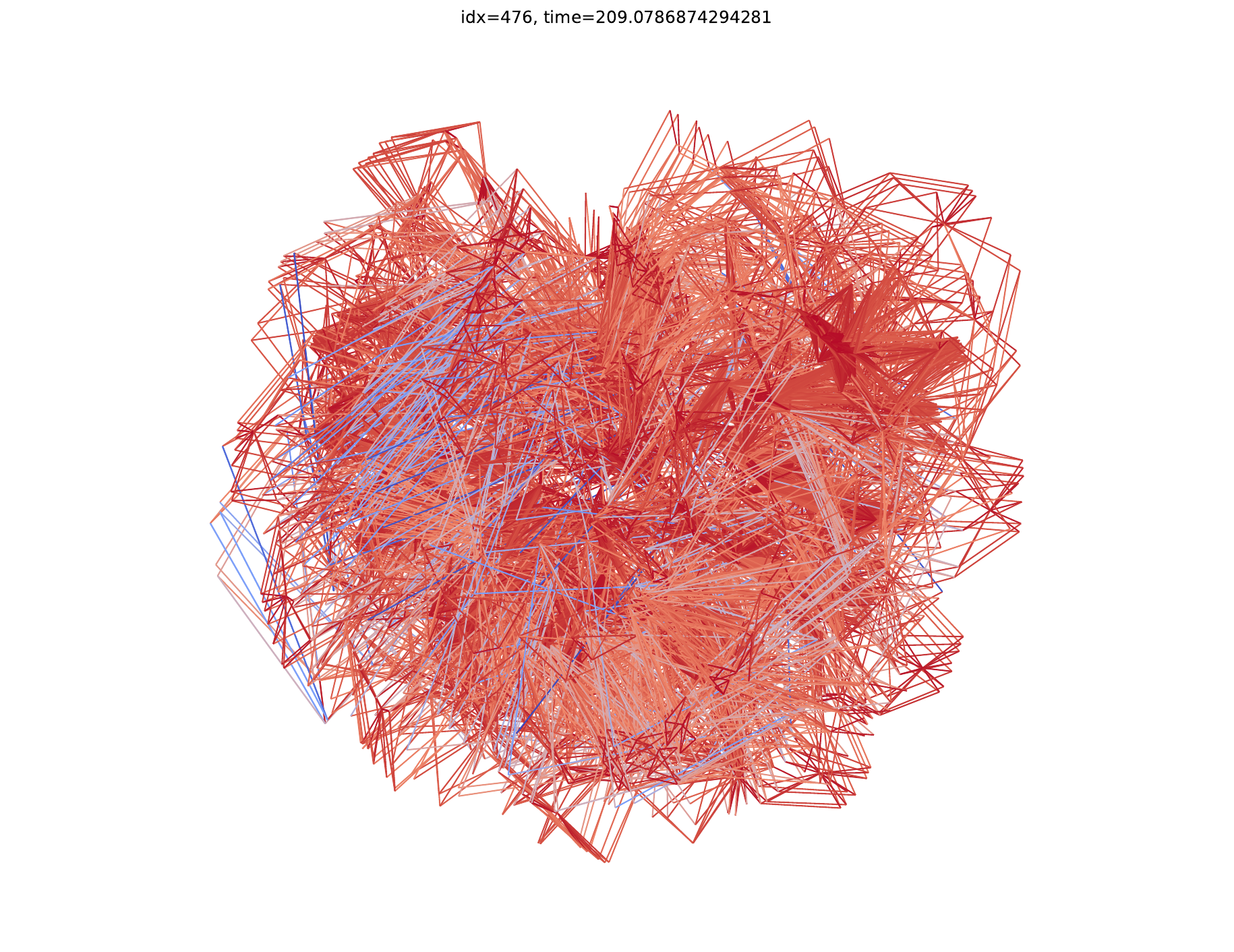} &
\imgcell{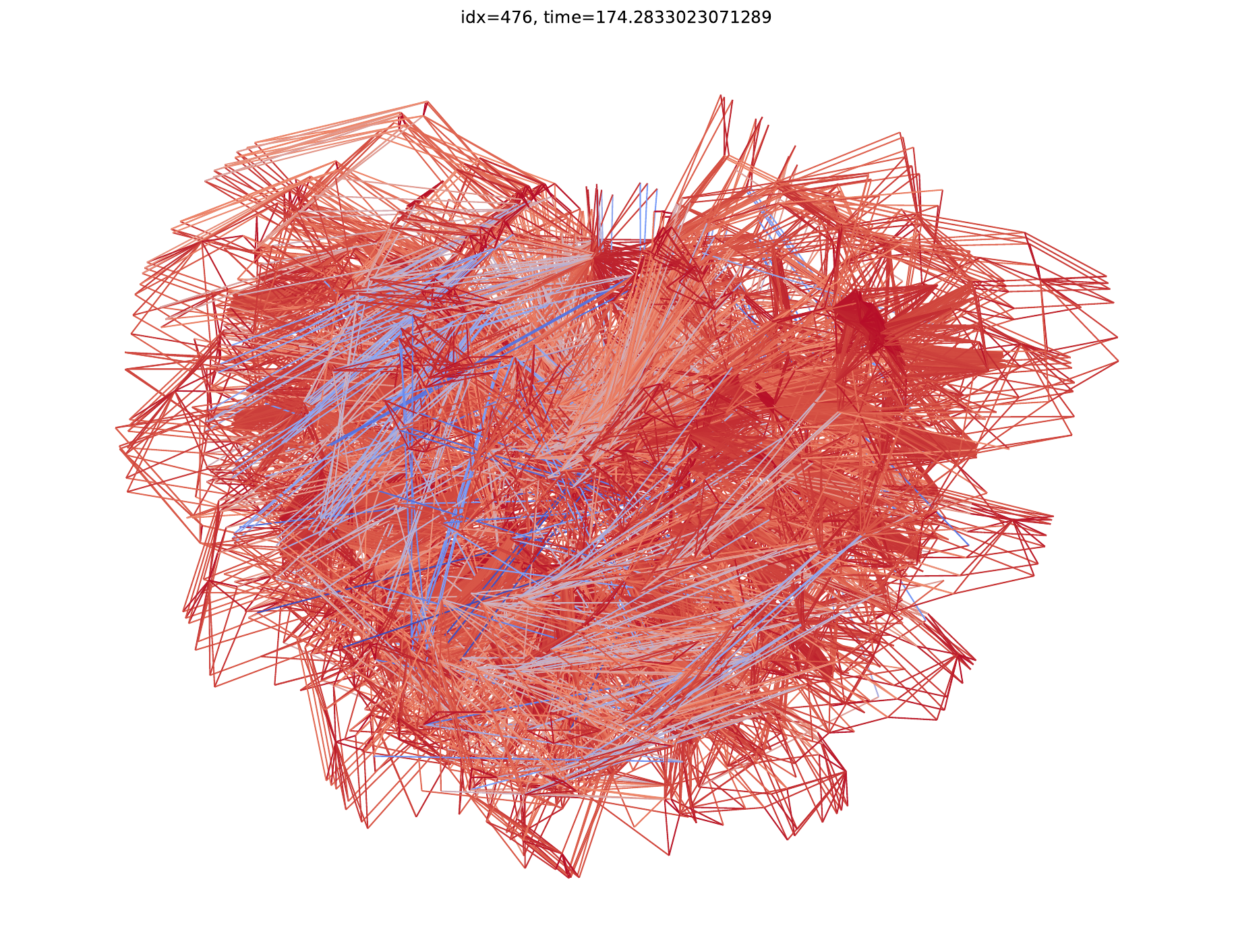} &
\imgcell{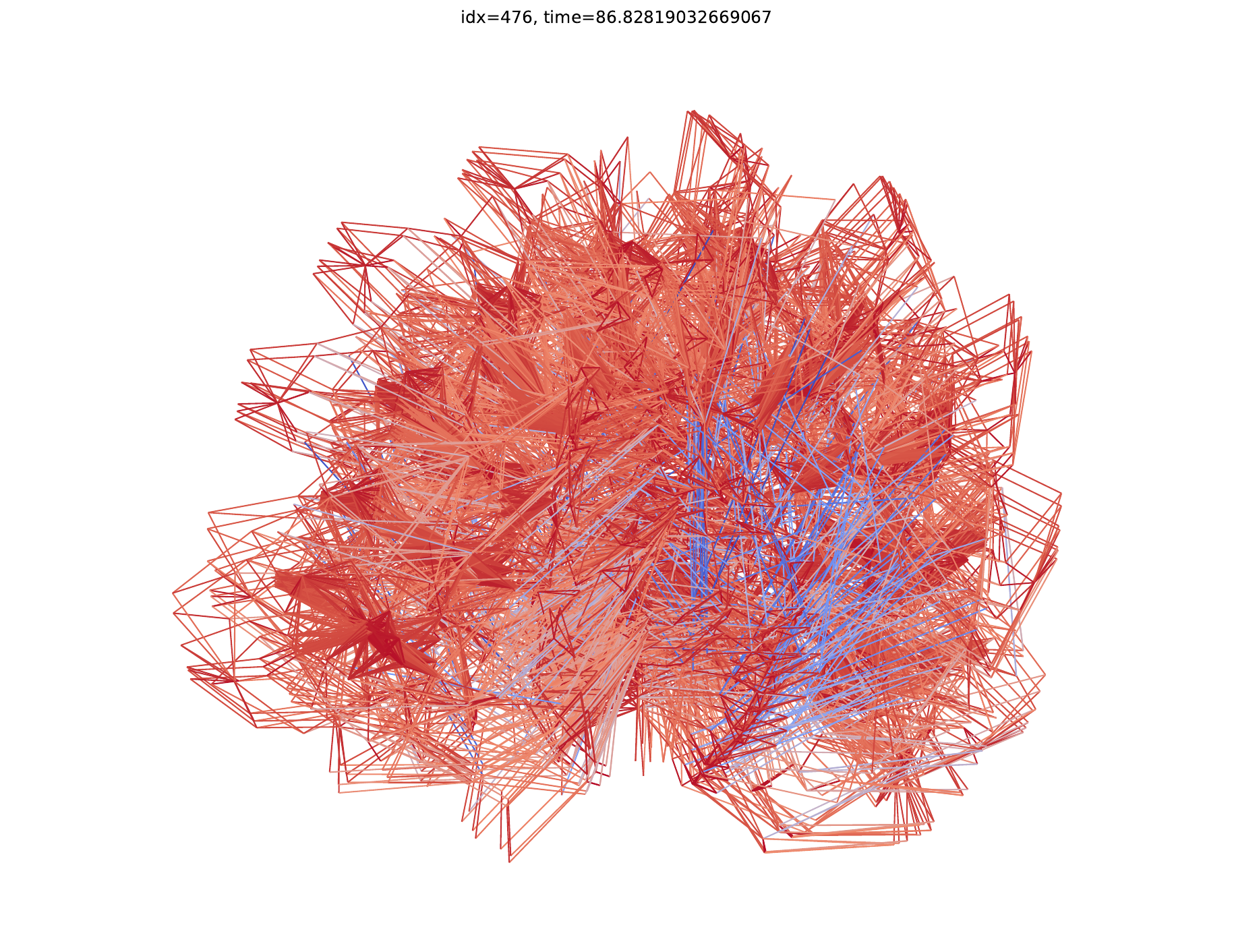} \\

&
t = 6.61s &
t = 78.93s &
t = 604.50s &
t = 28.81s &
t = 7200.00s &
t = 20.17s &
t = 29.90s &
t = 22.33s &
t = 19.05s &
t = 20.08s &
t = 27.28s &
t = 26.60s \\

\end{tabular}
\captionof{figure}[]{The qualitative evaluation of 7 \modelName\ models by comparing with 5 competitive and representative benchmarks. All the graphs presented above are unseen during the training phase of \modelName. The name of the graphs with the number of nodes $N$ and the number of edges $M$ is presented in the row header. For each layout, the computation time $t$ (without including the pre-processing time) on the CPU is computed and reported in seconds. }
\label{fig:more-vis-result2}
\end{table*}


\begin{table*}[ht!]
\setlength{\tabcolsep}{0pt}
\renewcommand{\arraystretch}{0}
\fontsize{6}{6}\selectfont
\centering
\begin{tabular}{ c|ccccc|ccccccc }
    \bfseries{\thead{Graph}} & \multicolumn{5}{c|}{\thead{Benchmark Methods}} & \multicolumn{6}{c}{\thead{SmartGD}}\\
    & \bfseries{SGD2}
    & \bfseries{PMDS}
    & \bfseries{FA2}
    & \bfseries{DeepGD}
    & \makecell{\bfseries GD2\\\relax[Stress+Xing]}
    & \makecell{\bfseries SmartGD\\\relax[Stress]}
    & \makecell{\bfseries SmartGD\\\relax[Xing]}
    & \makecell{\bfseries SmartGD\\\relax[Shape]}
    & \makecell{\bfseries SmartGD\\\relax[XAngle]}
    & \makecell{\bfseries SmartGD\\\relax[Stress+Xing]}
    & \makecell{\bfseries SmartGD\\\relax[Stress+XAngle]}
    & \makecell{\bfseries SmartGD\\\relax[7-Aesthetics]}
    \rule[-1ex]{0pt}{0ex} \\ \hline

\makecell{\bfseries grafo11364.38\\N = 16\\M = 19} &
\imgcell{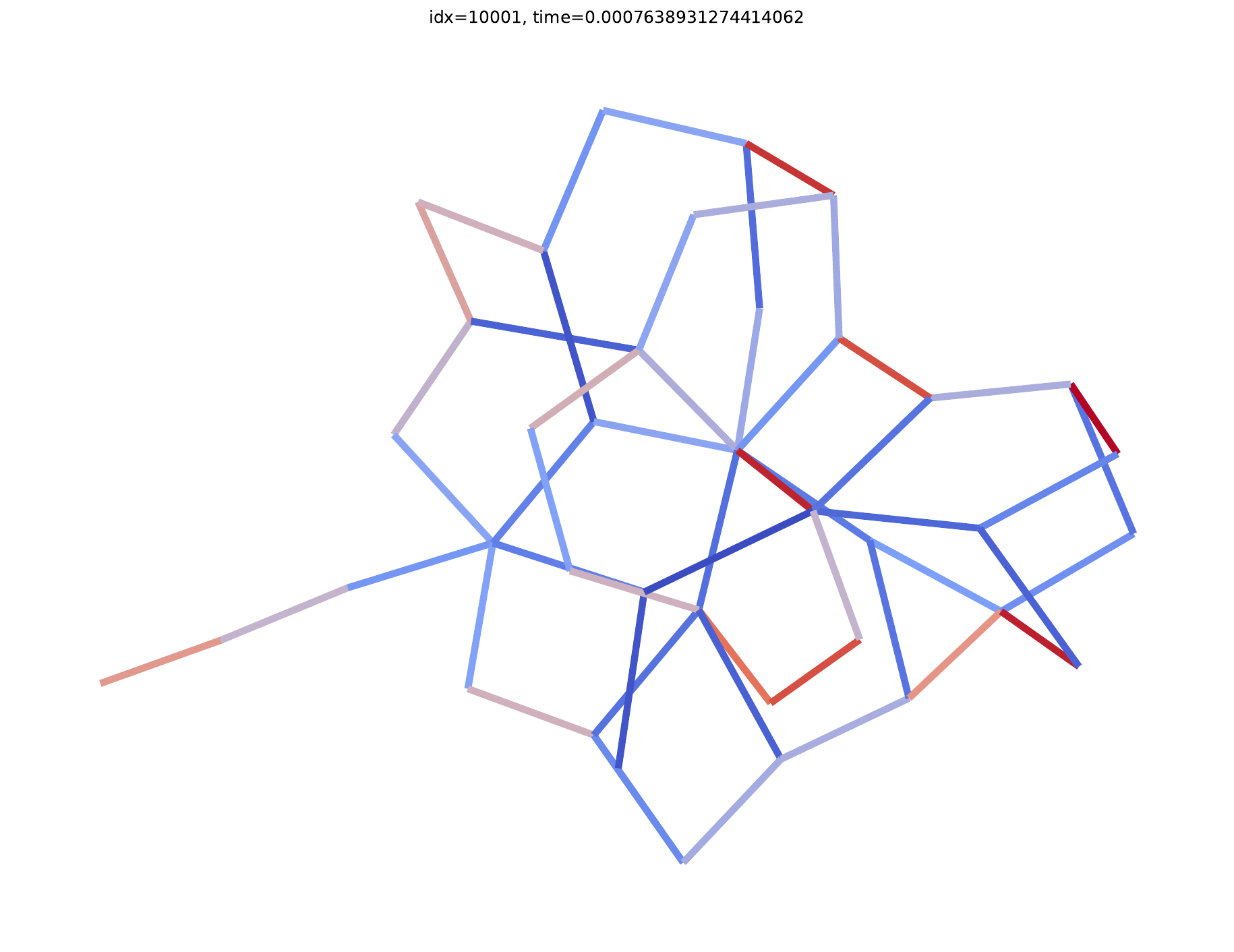} &
\imgcell{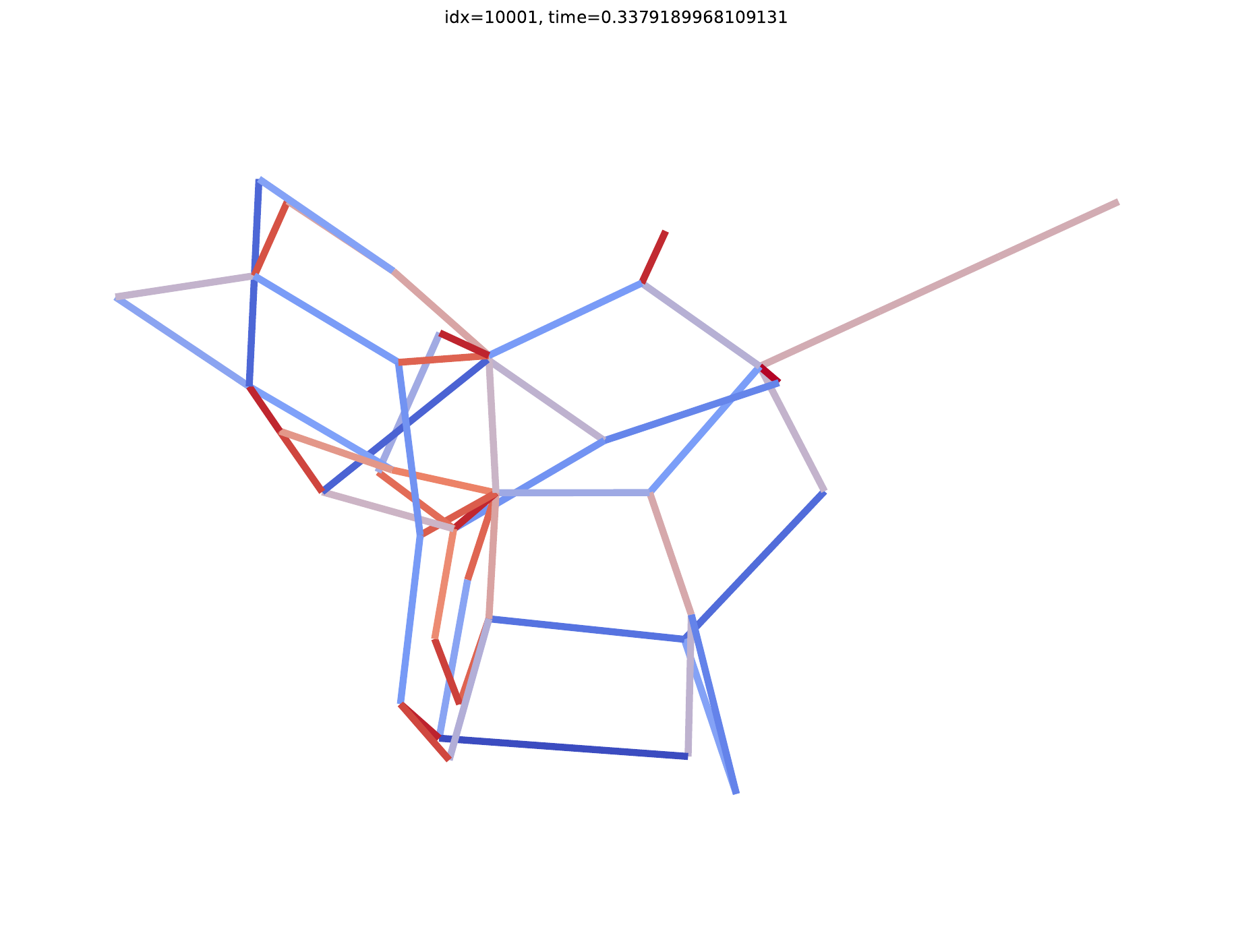} &
\imgcell{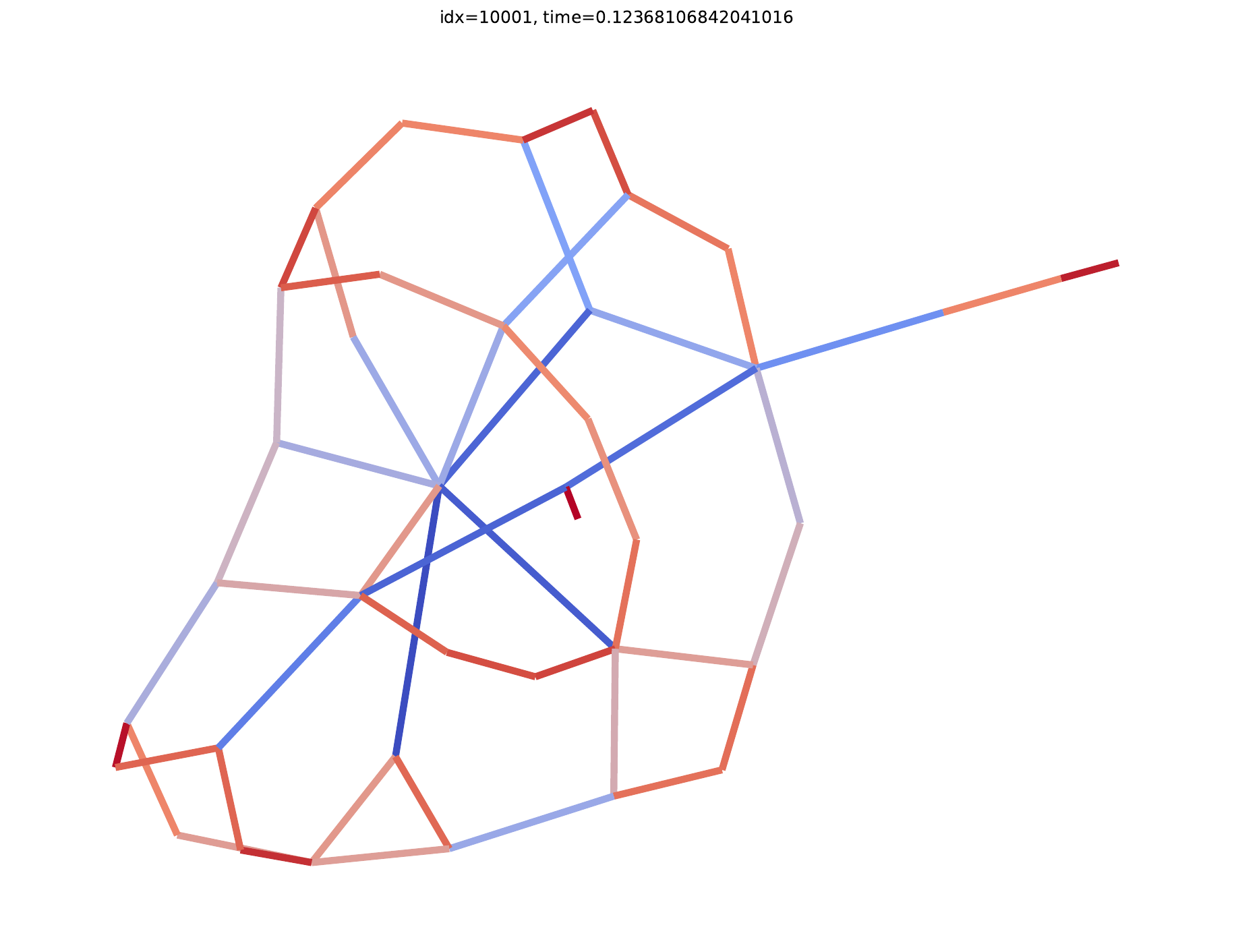} &
\imgcell{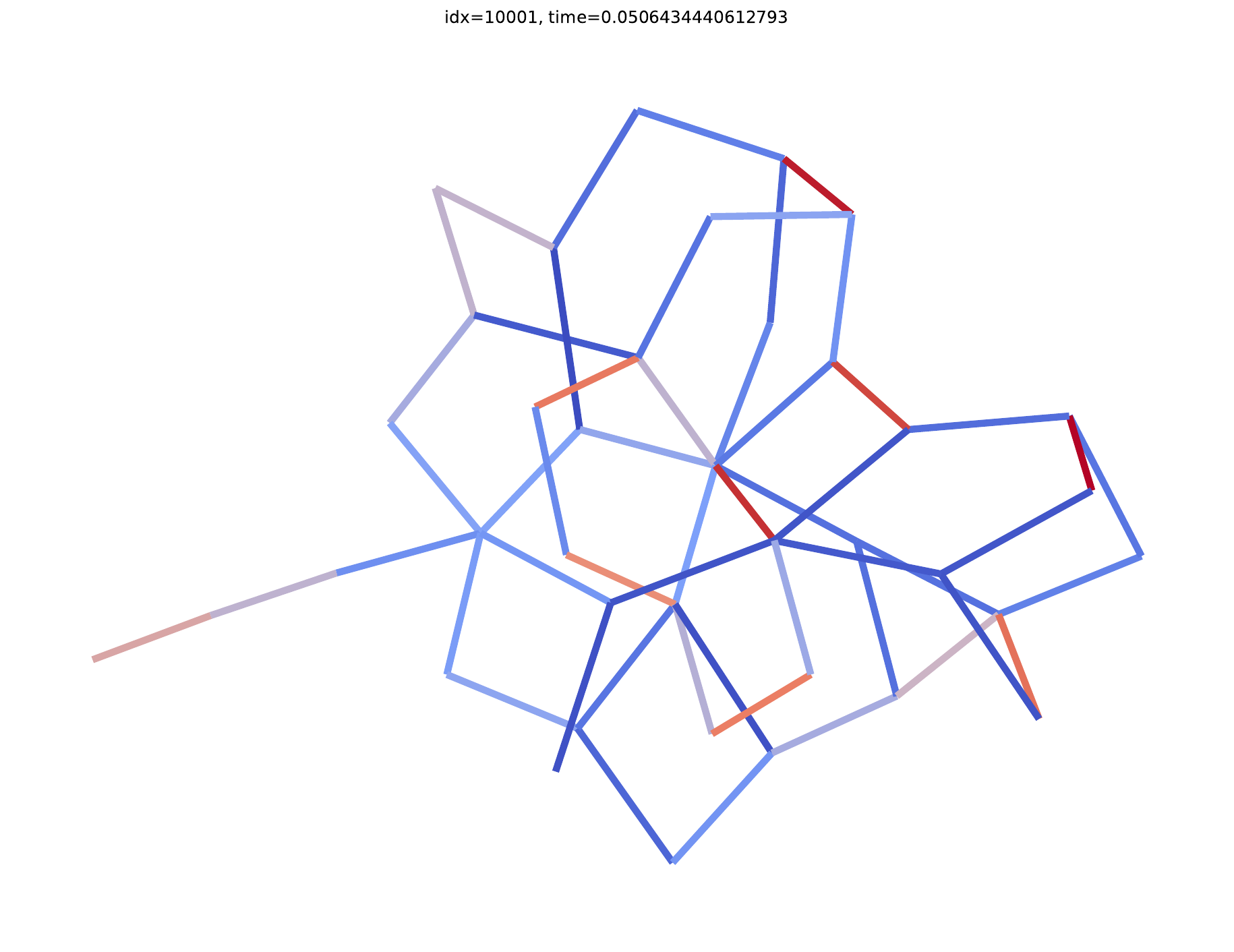} &
\imgcell{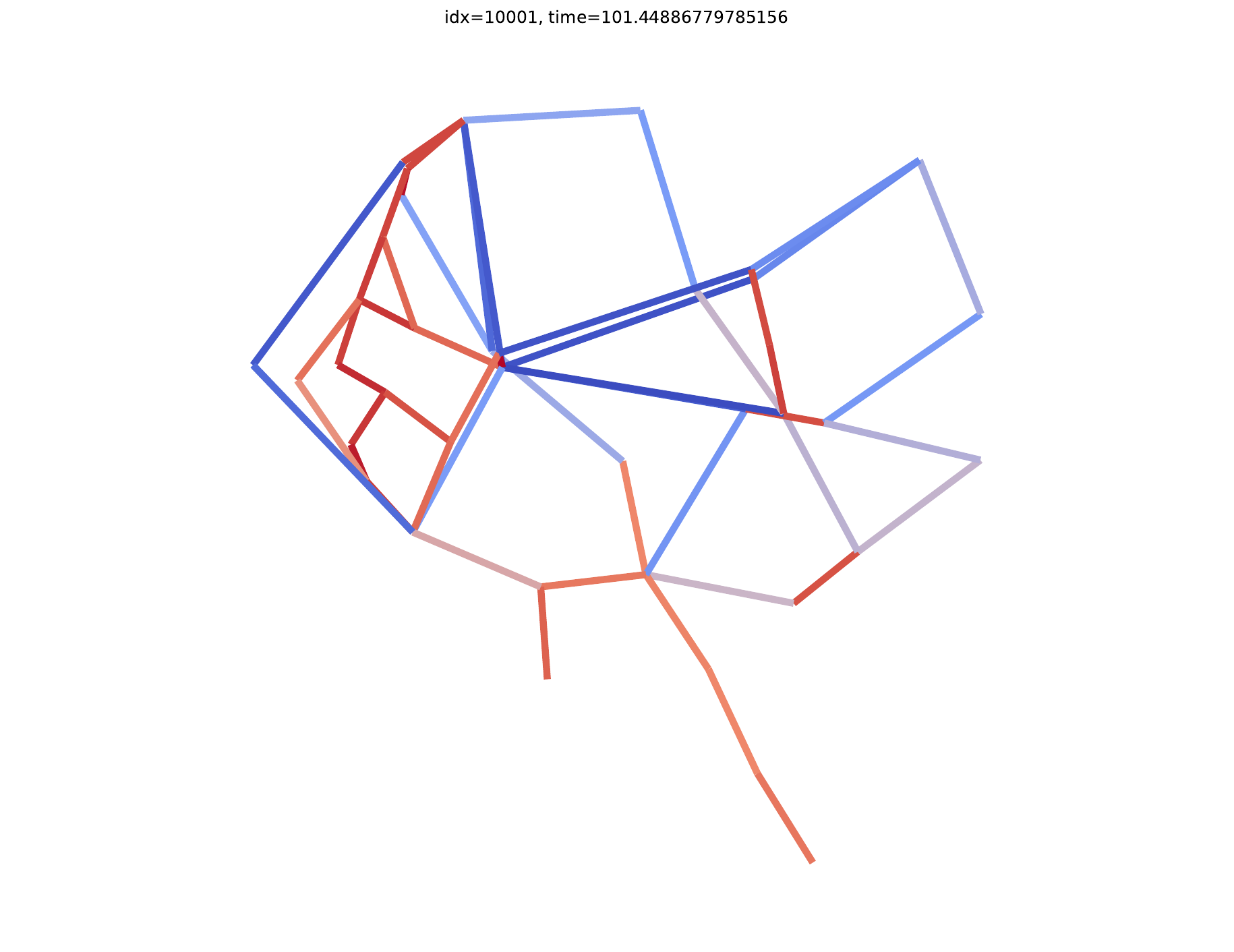} &
\imgcell{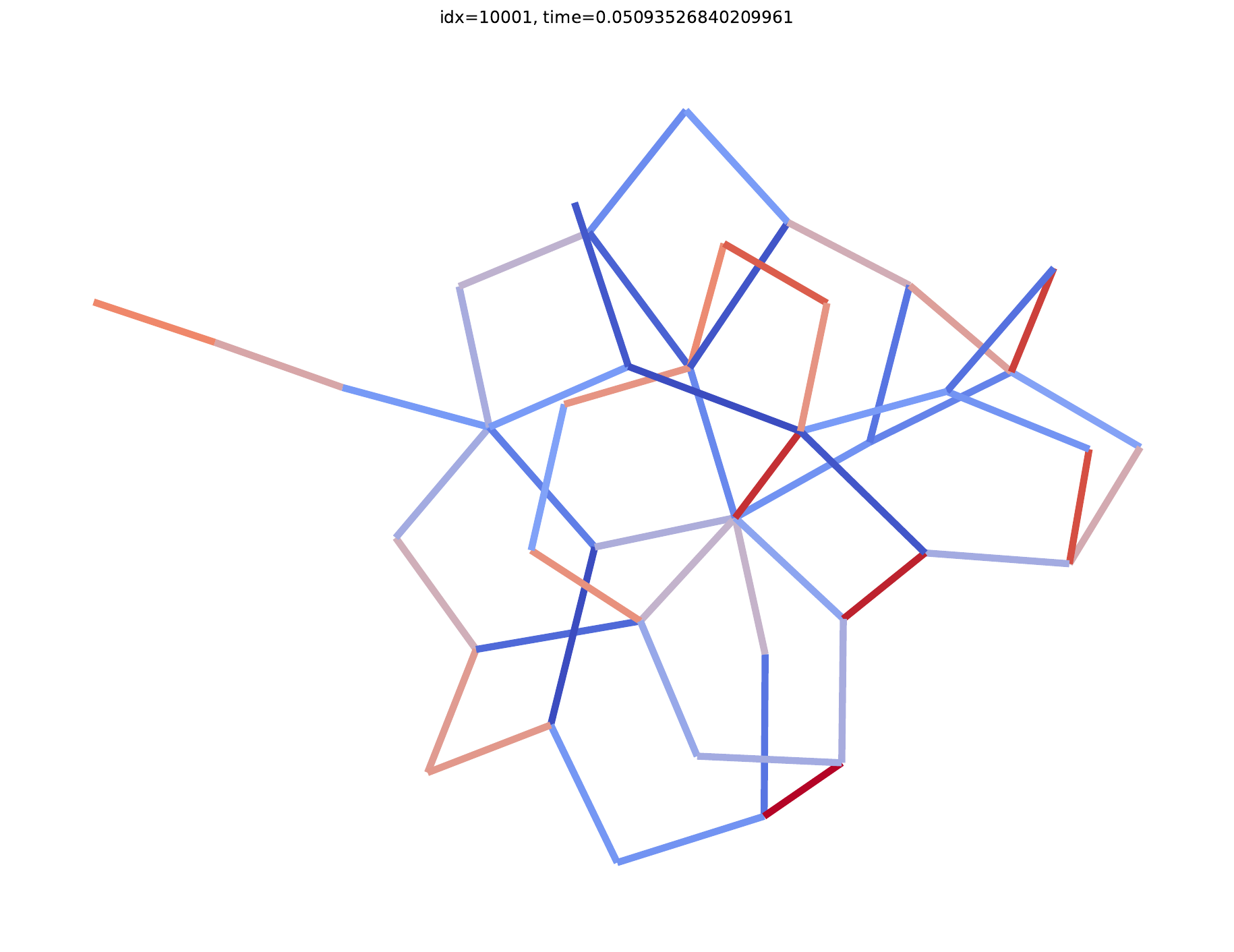} &
\imgcell{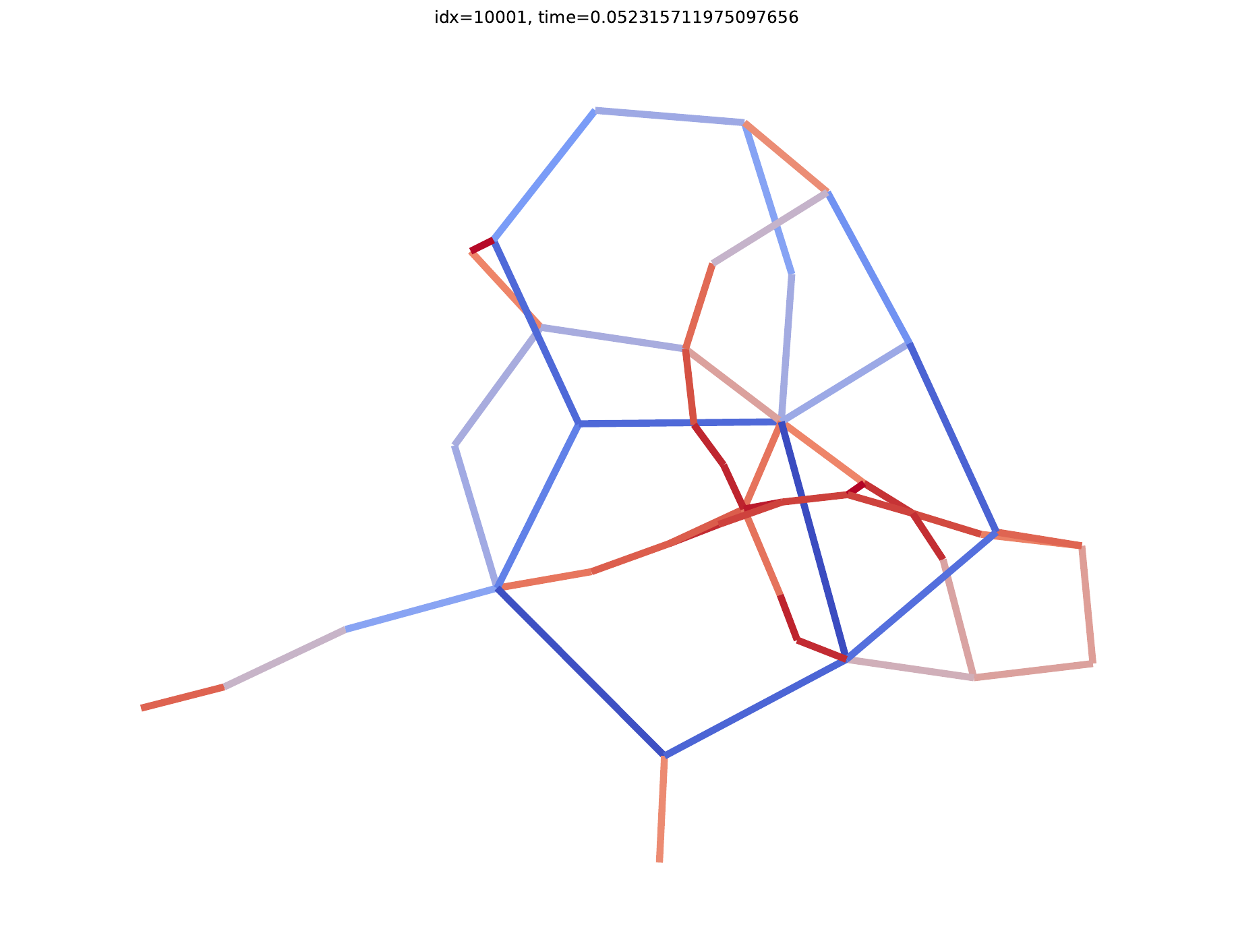} &
\imgcell{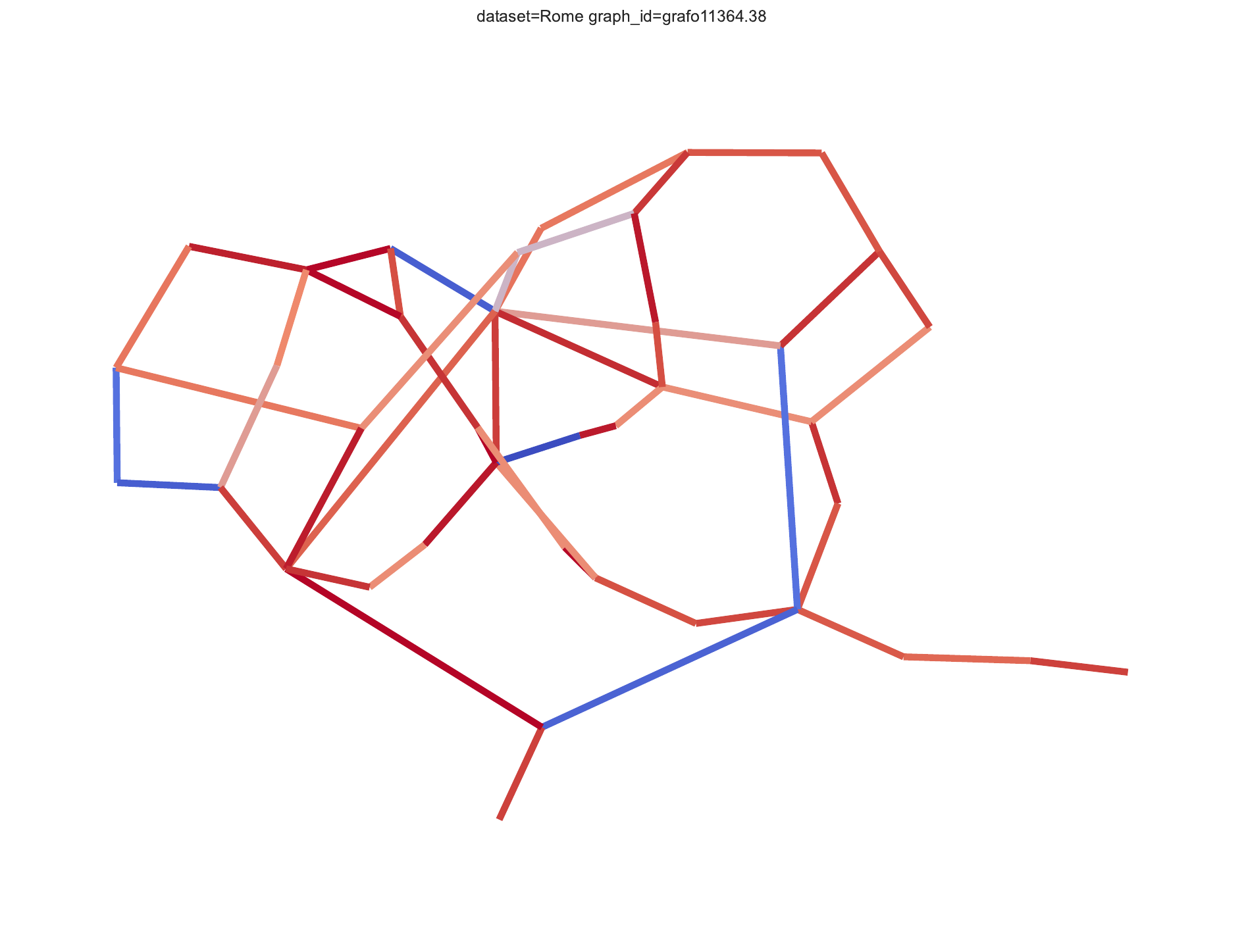} &
\imgcell{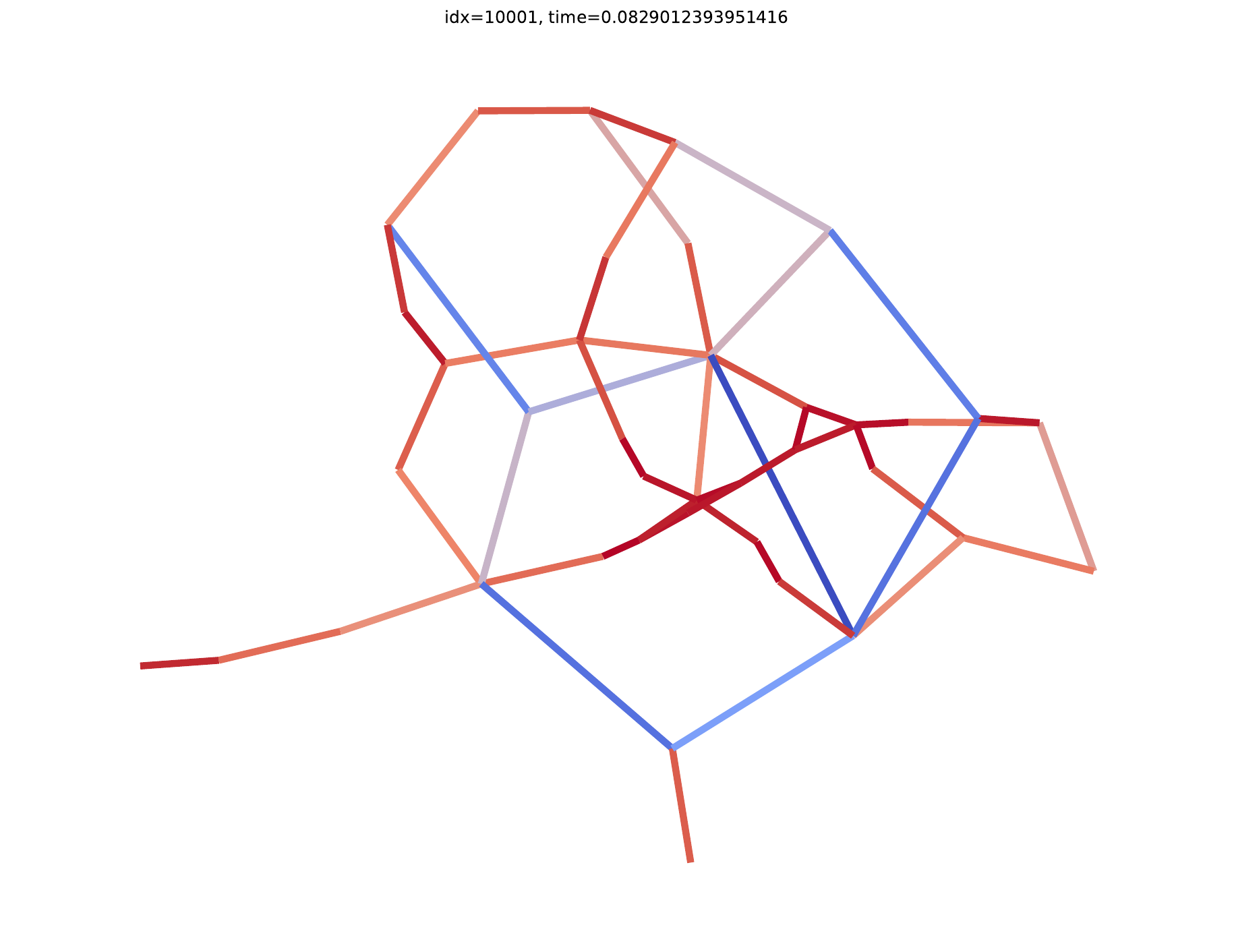} &
\imgcell{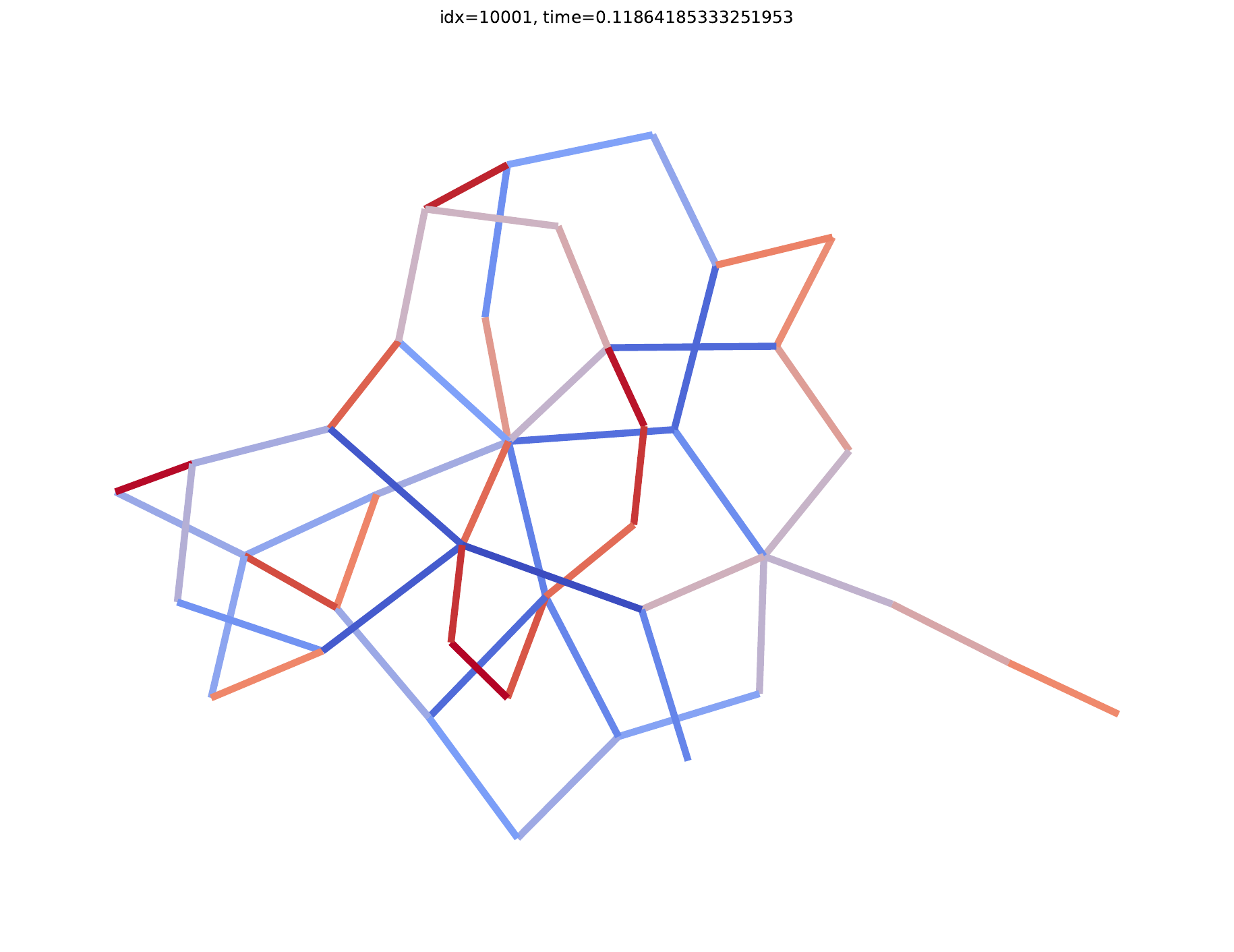} &
\imgcell{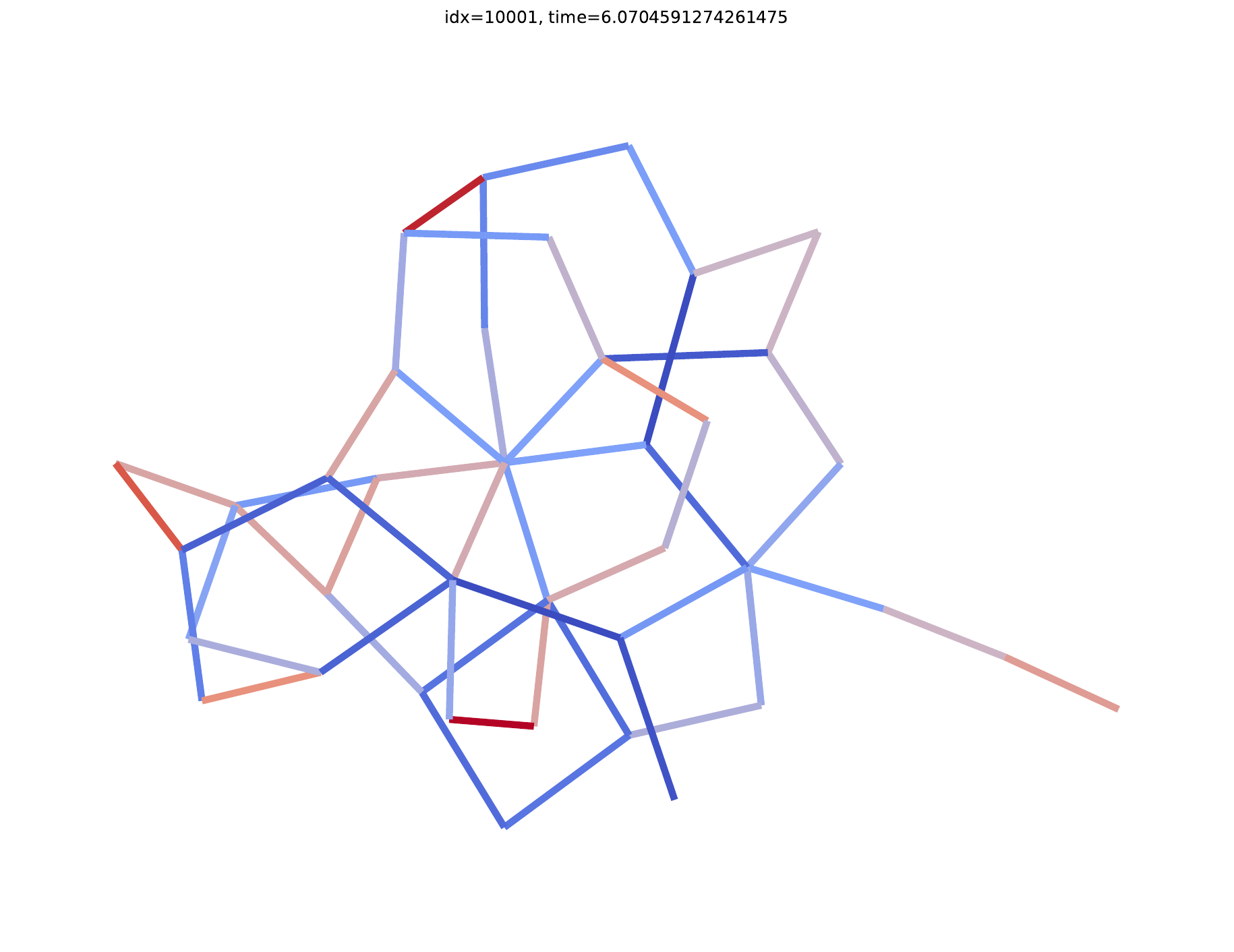} &
\imgcell{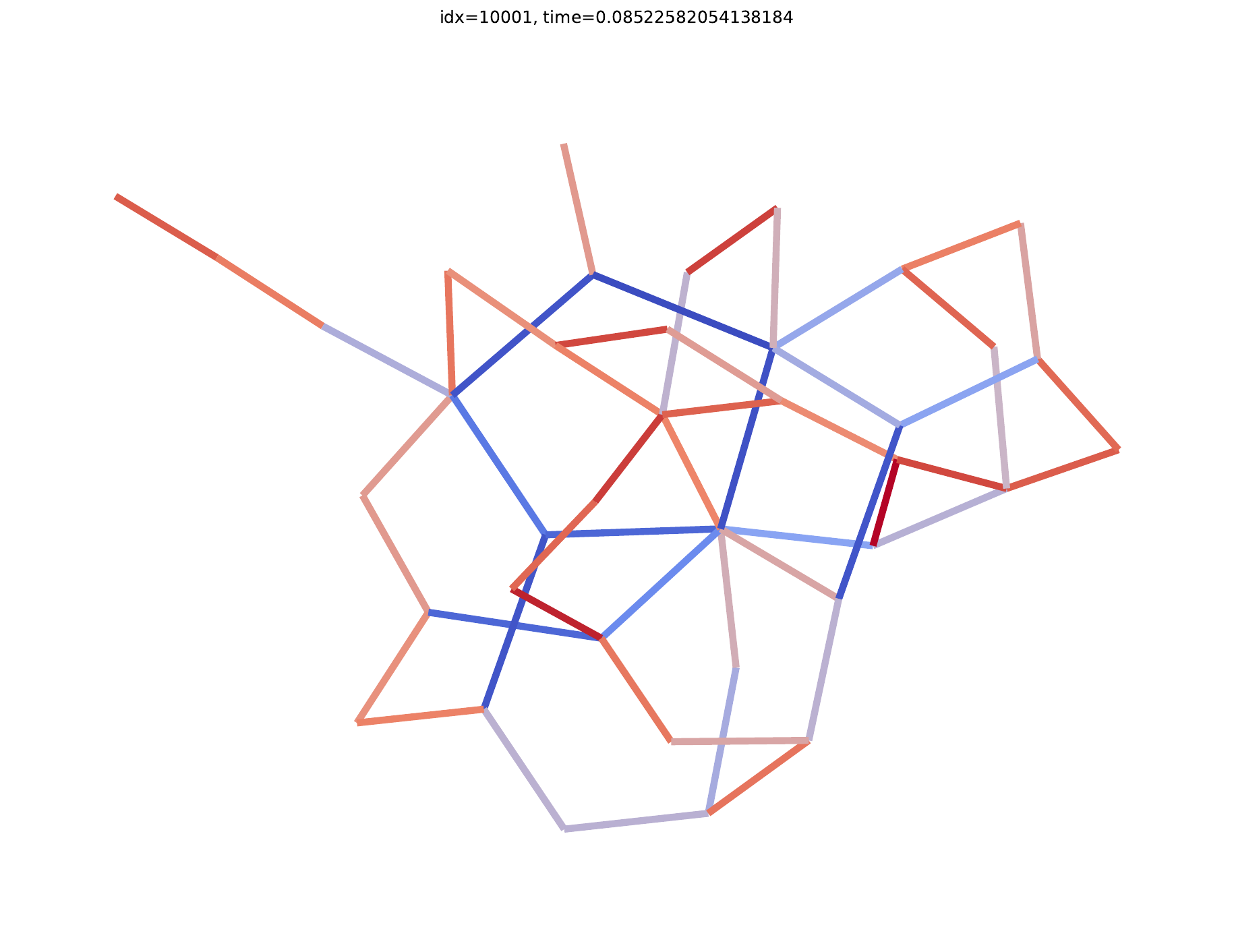} \\

&
t = 0.00s &
t = 0.34s &
t = 0.12s &
t = 0.05s &
t = 101.45s &
t = 0.05s &
t = 0.05s &
t = 0.09s &
t = 0.08s &
t = 0.12s &
t = 0.07s &
t = 0.09s \\

\makecell{\bfseries grafo2102.34\\N = 40\\M = 52} &
\imgcell{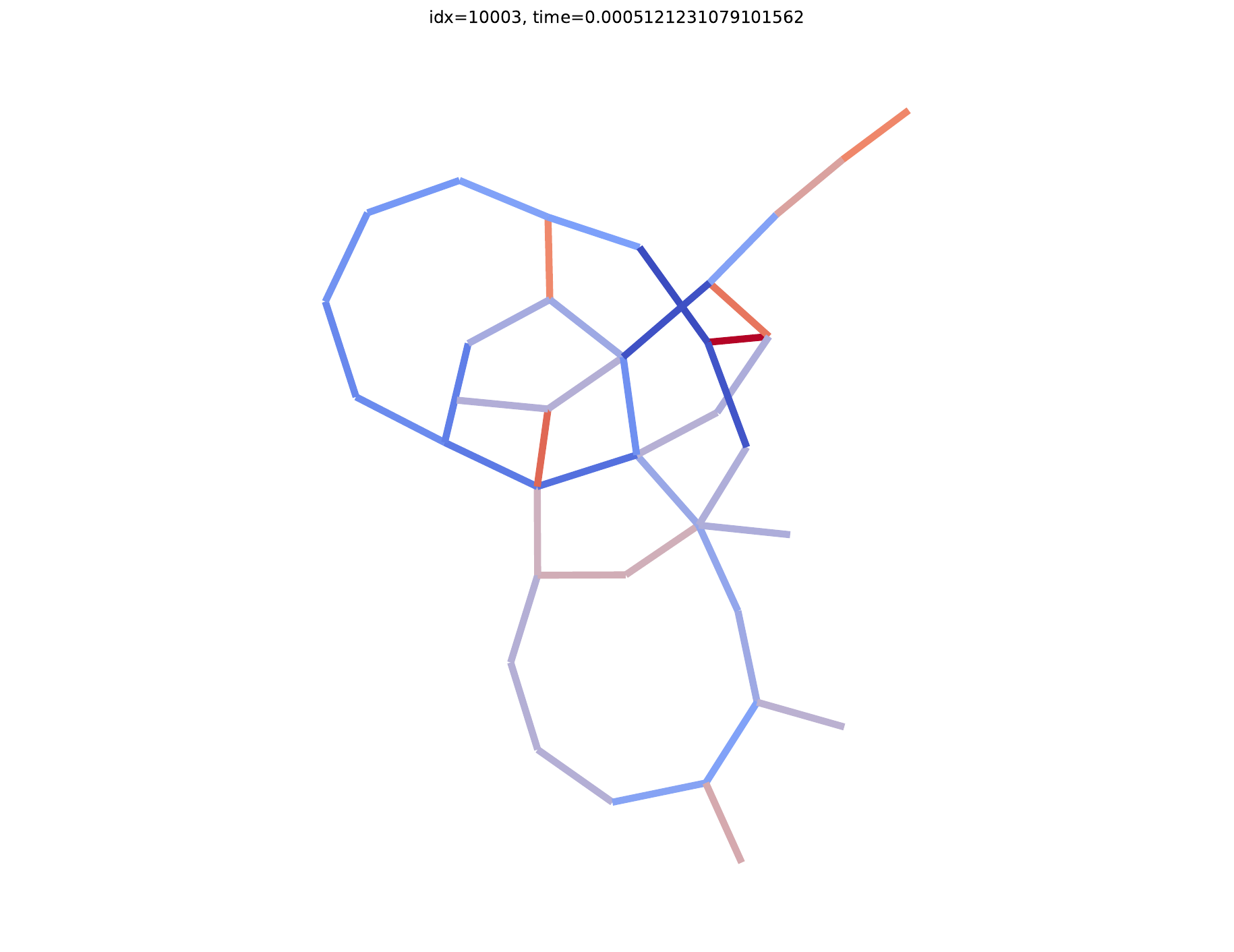} &
\imgcell{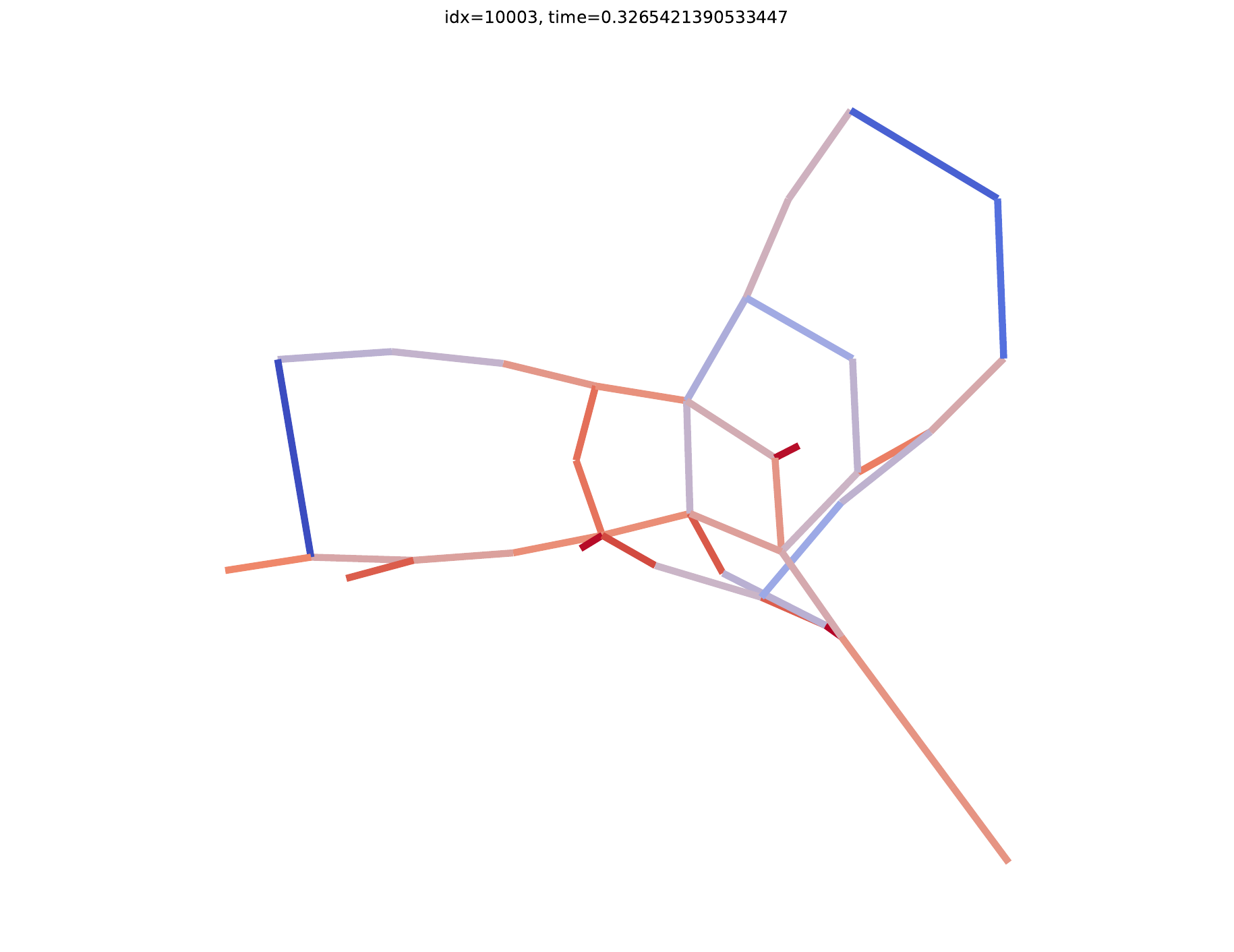} &
\imgcell{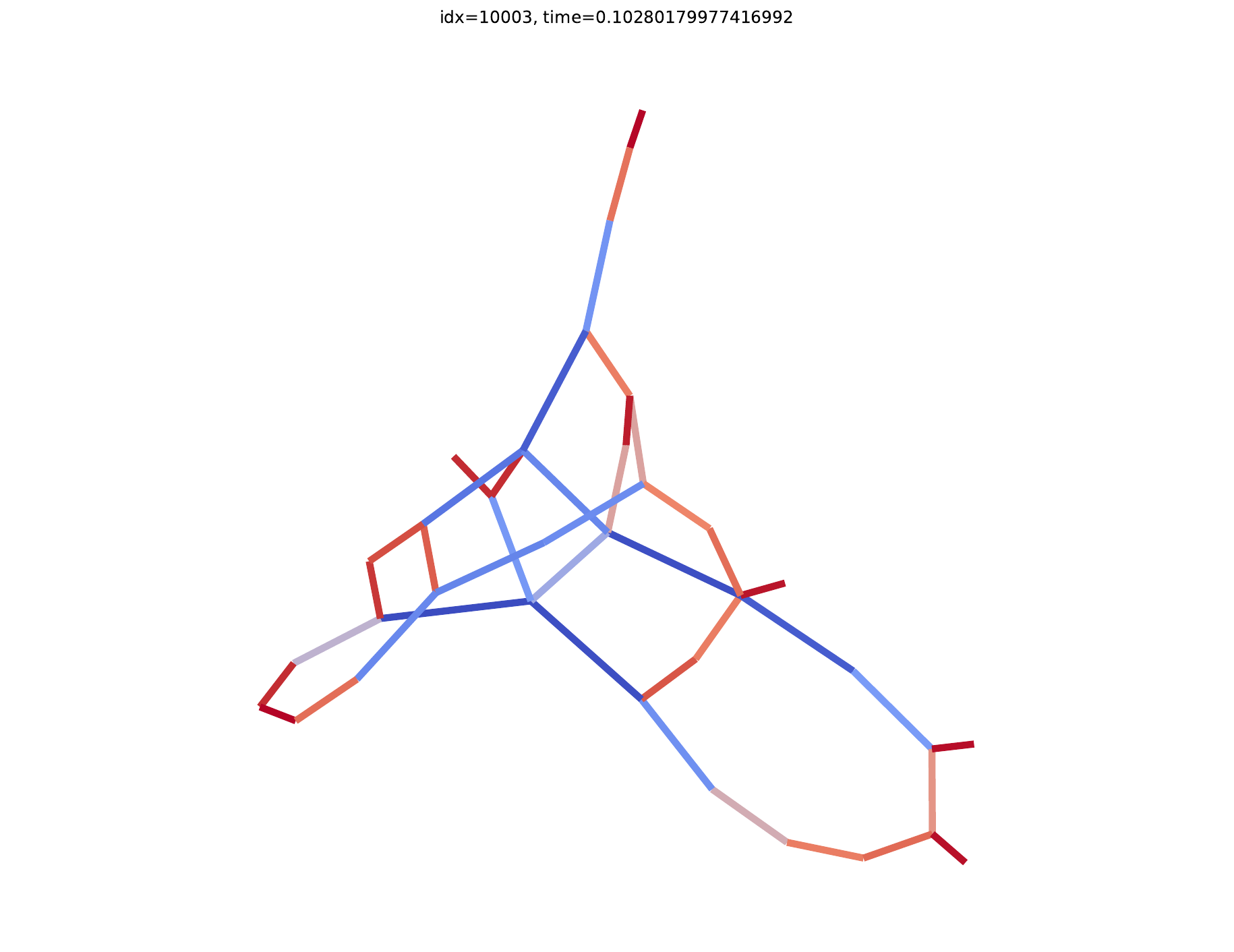} &
\imgcell{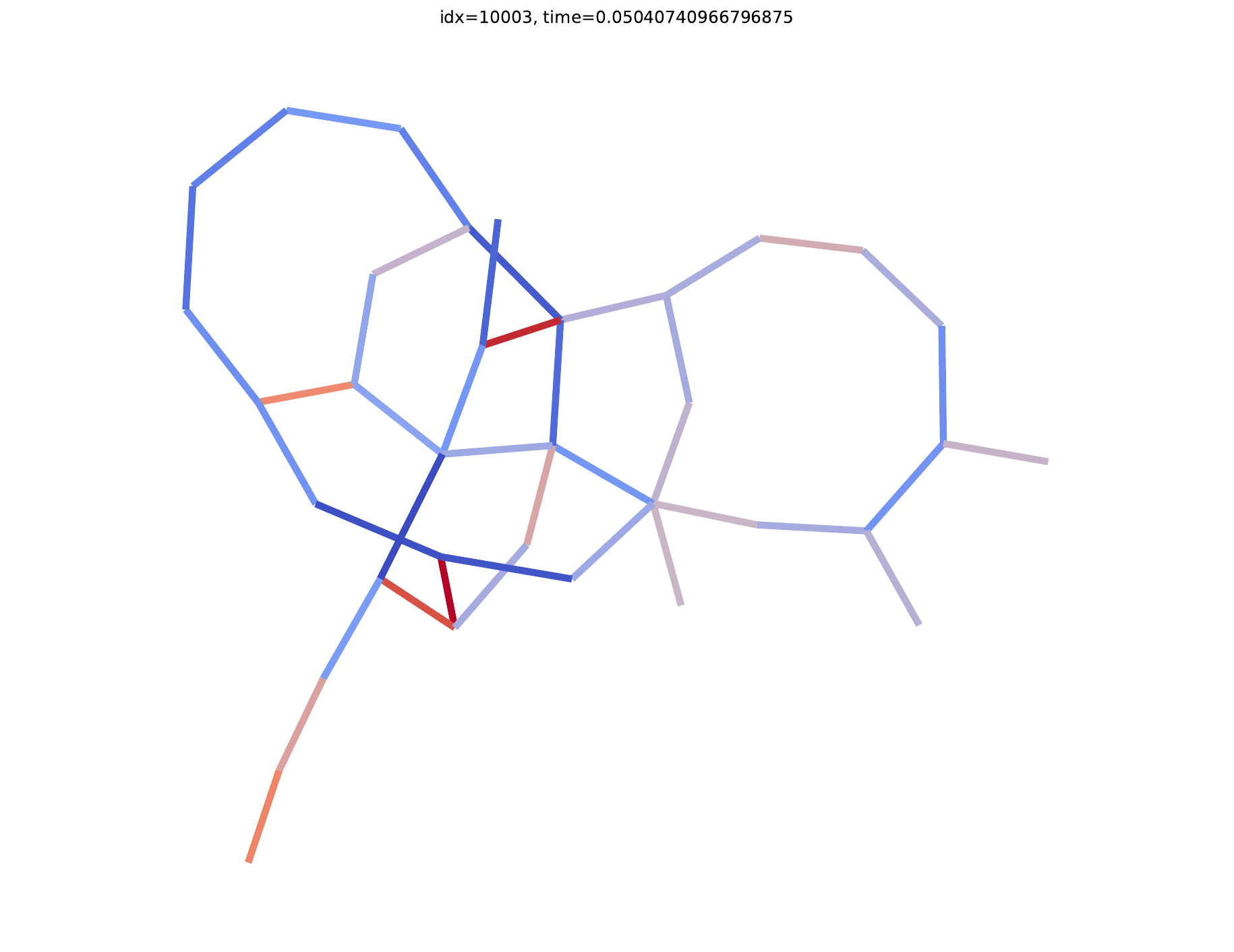} &
\imgcell{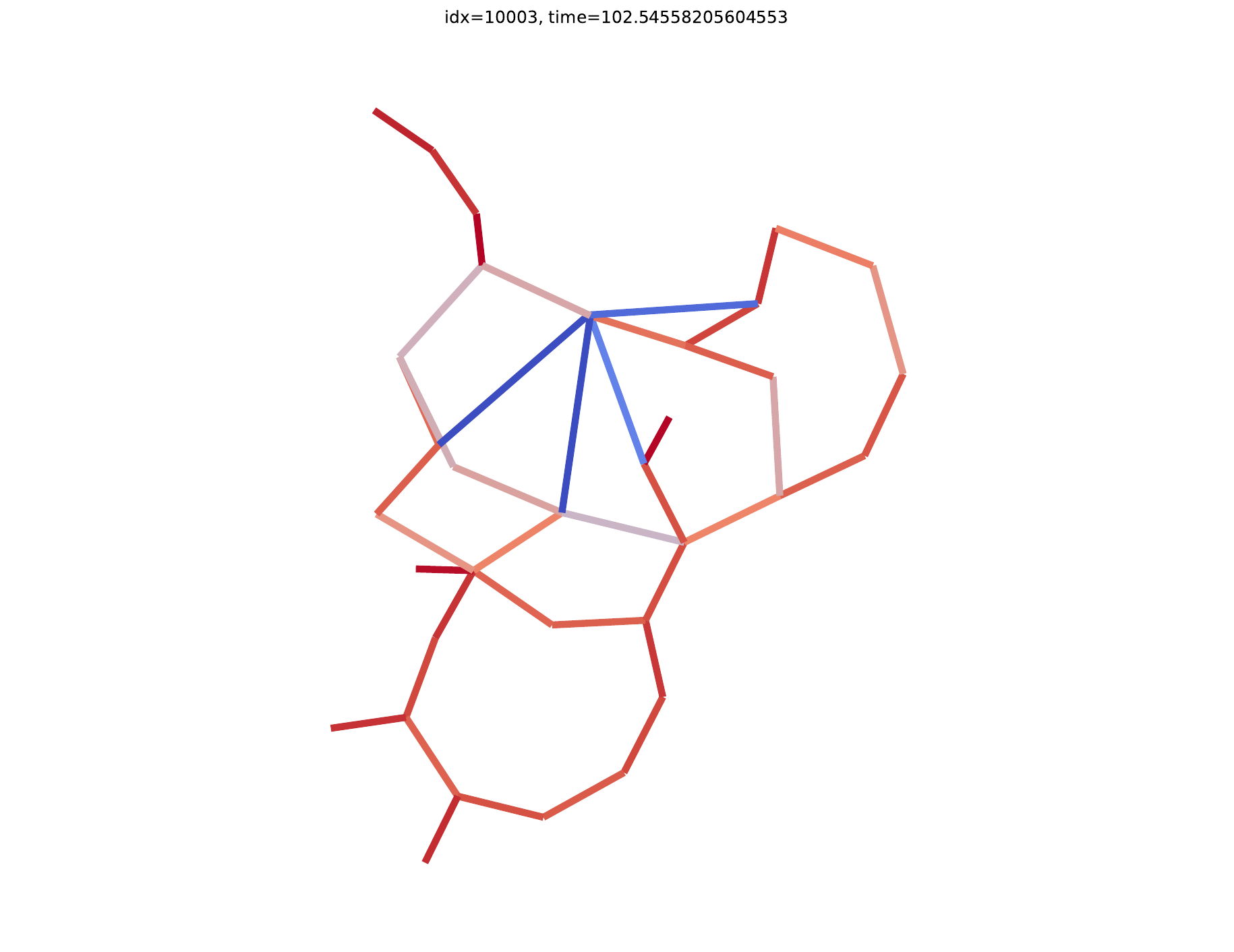} &
\imgcell{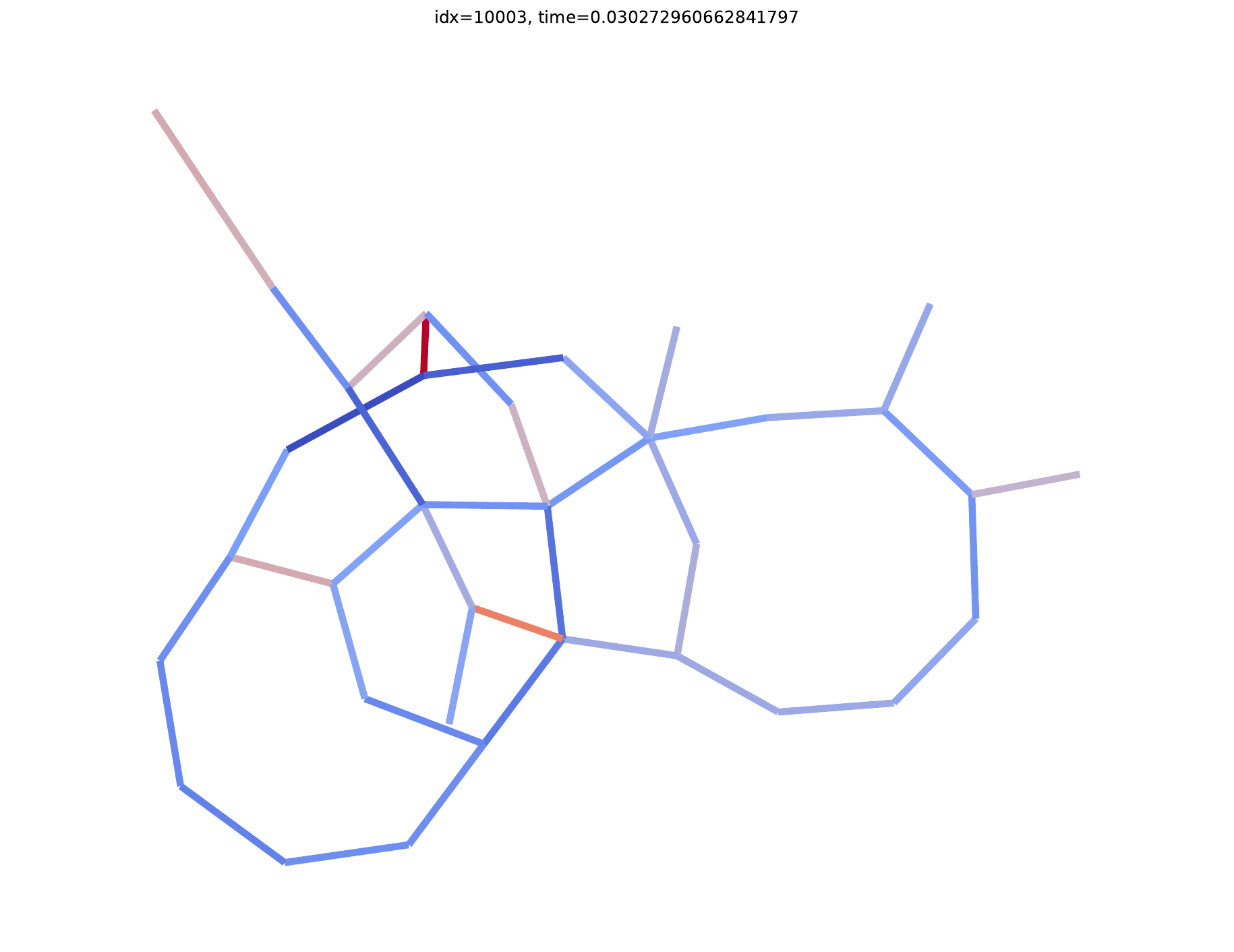} &
\imgcell{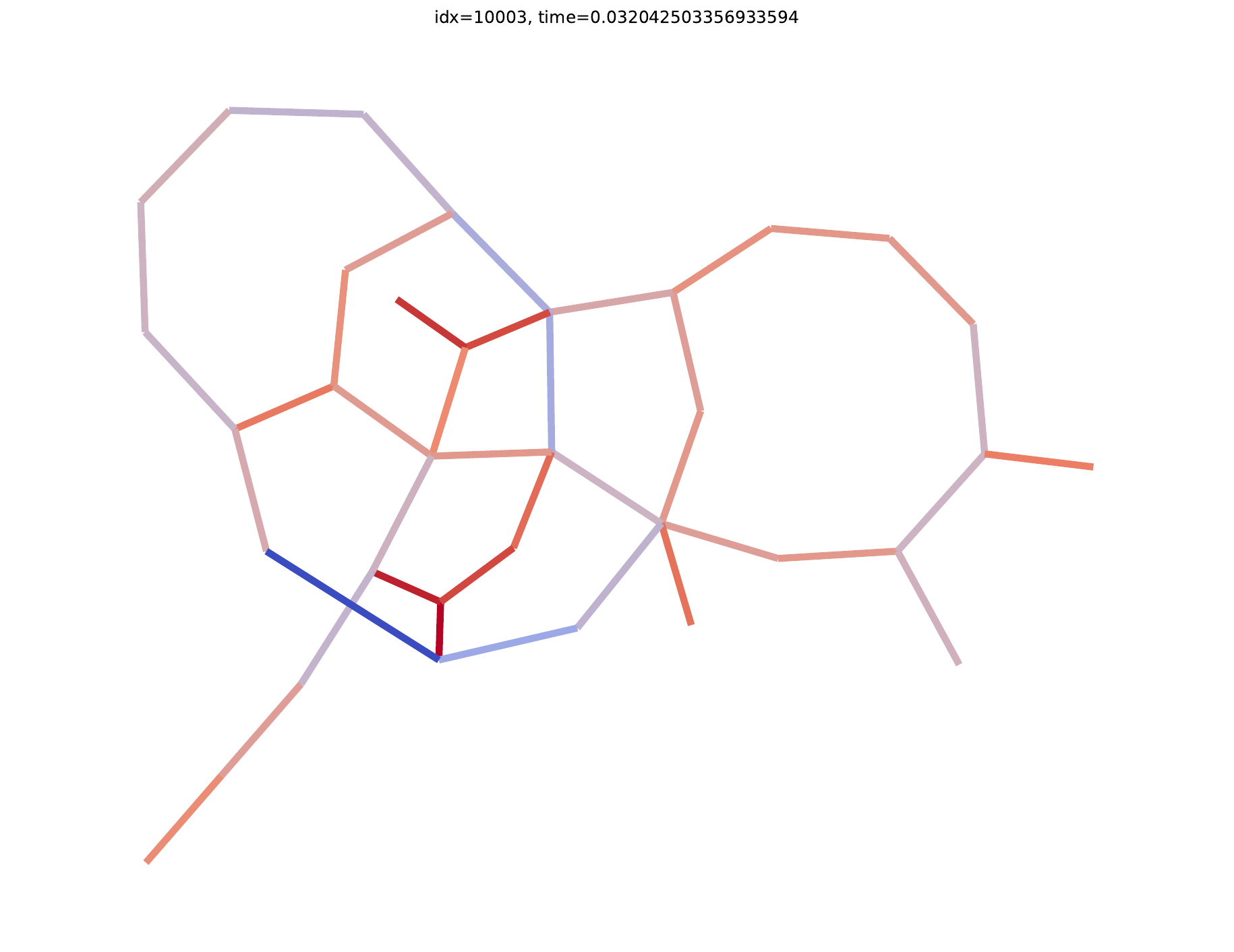} &
\imgcell{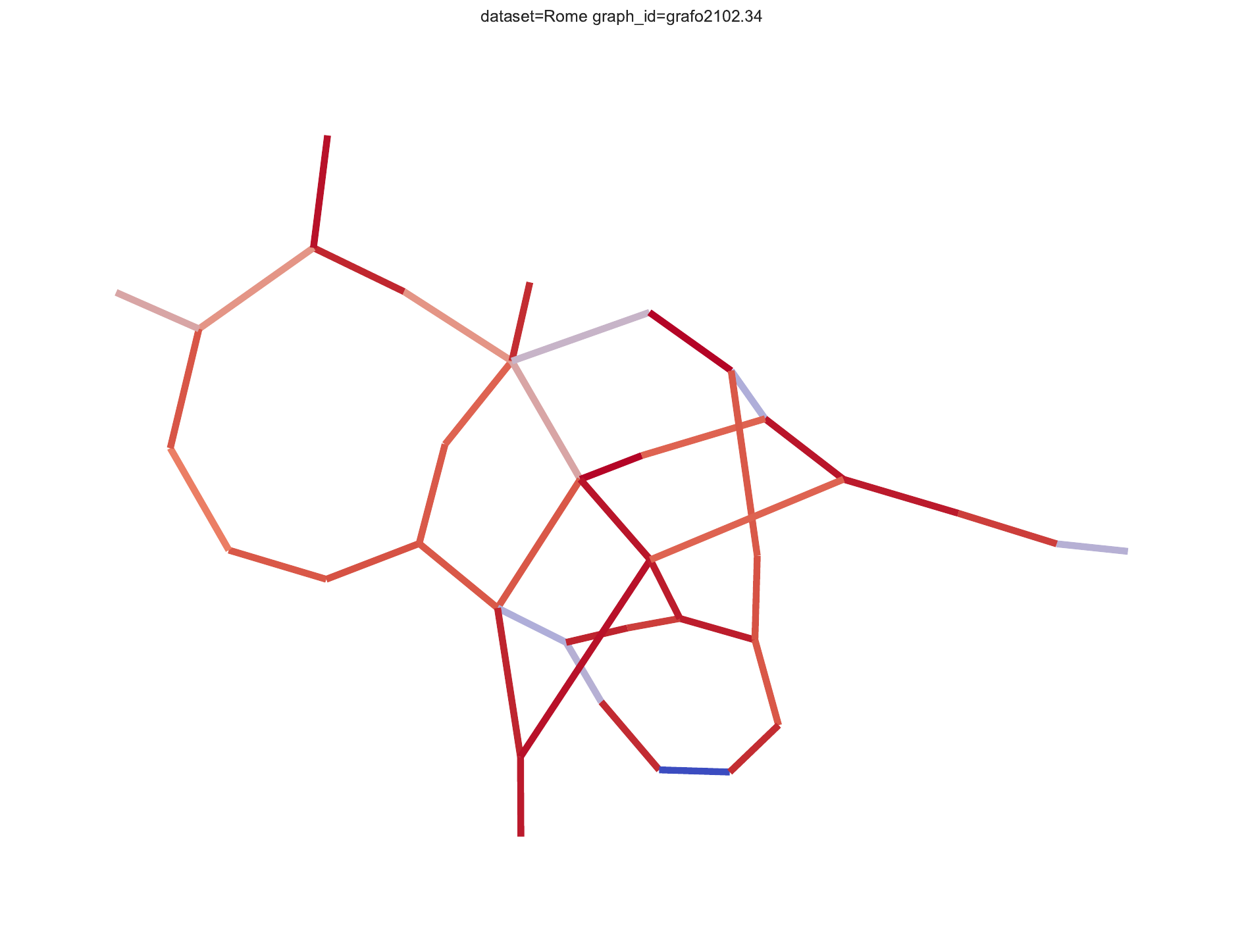} &
\imgcell{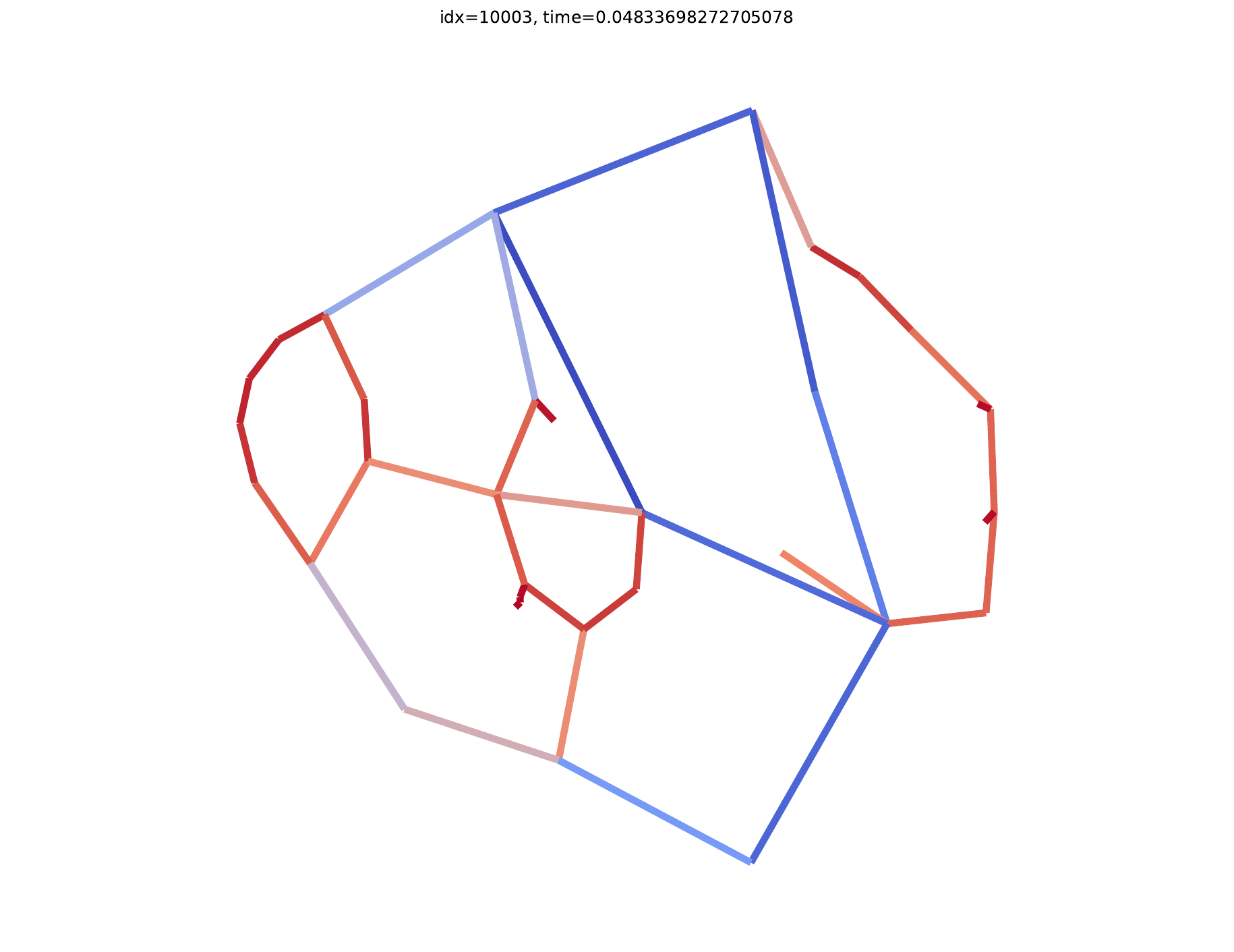} &
\imgcell{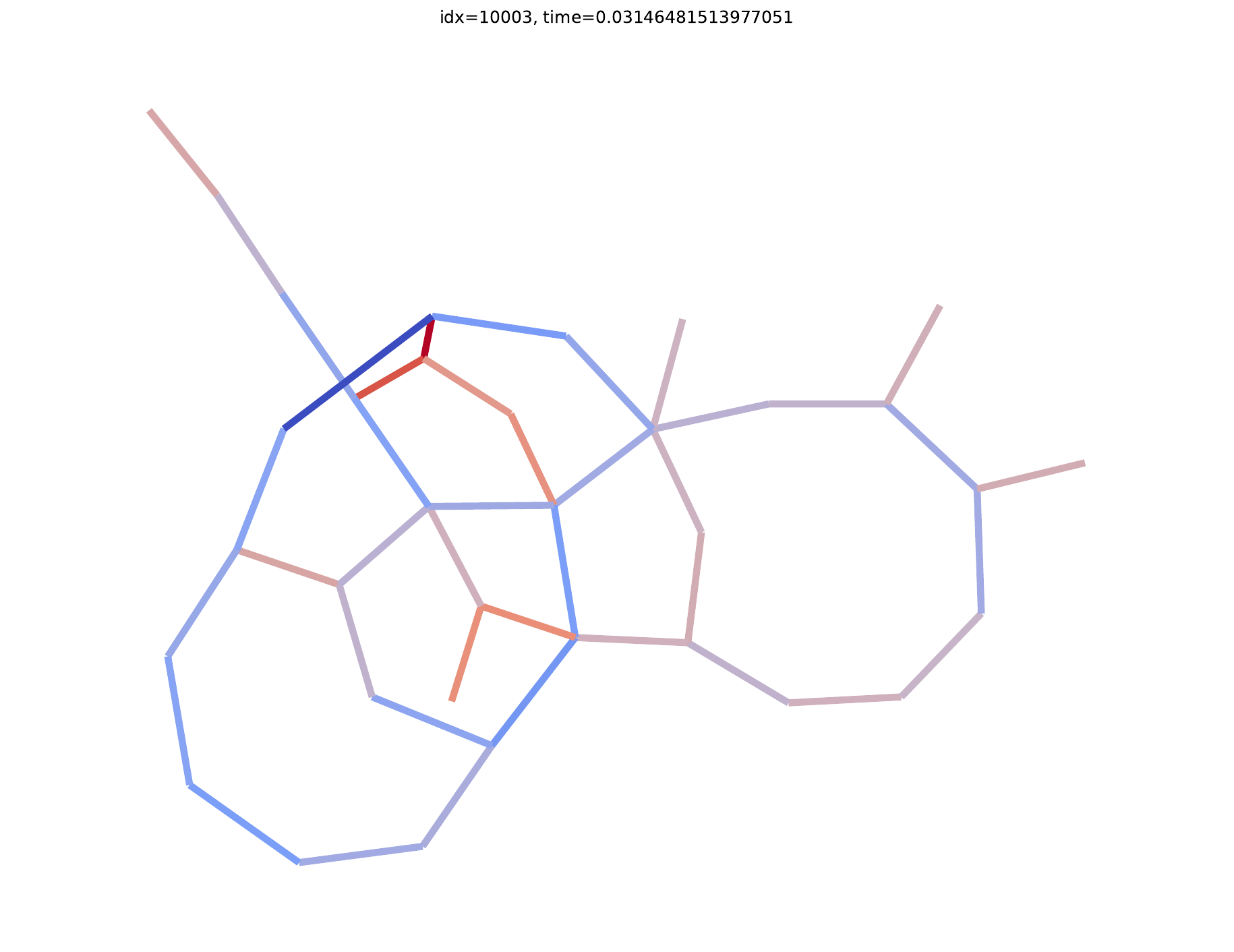} &
\imgcell{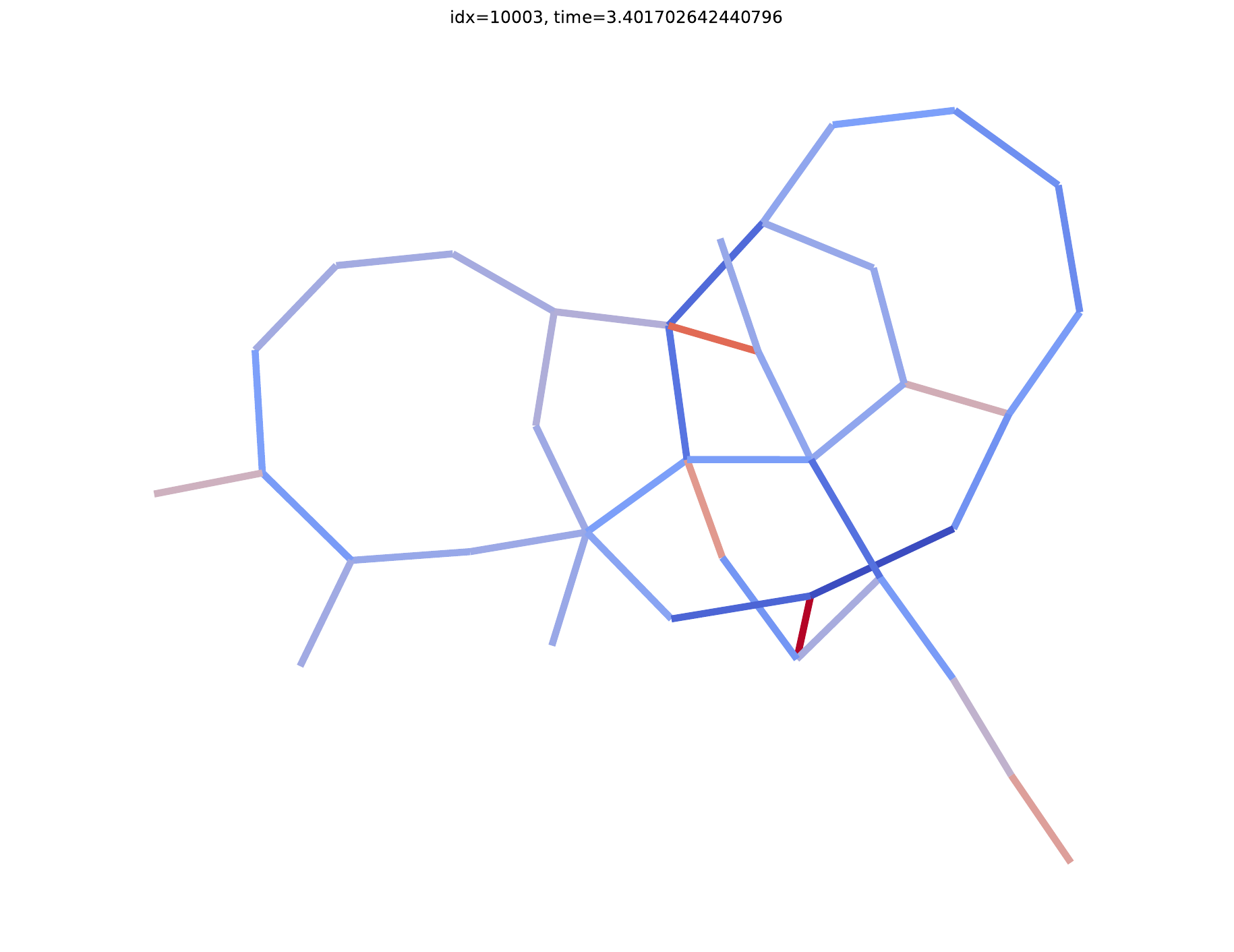} &
\imgcell{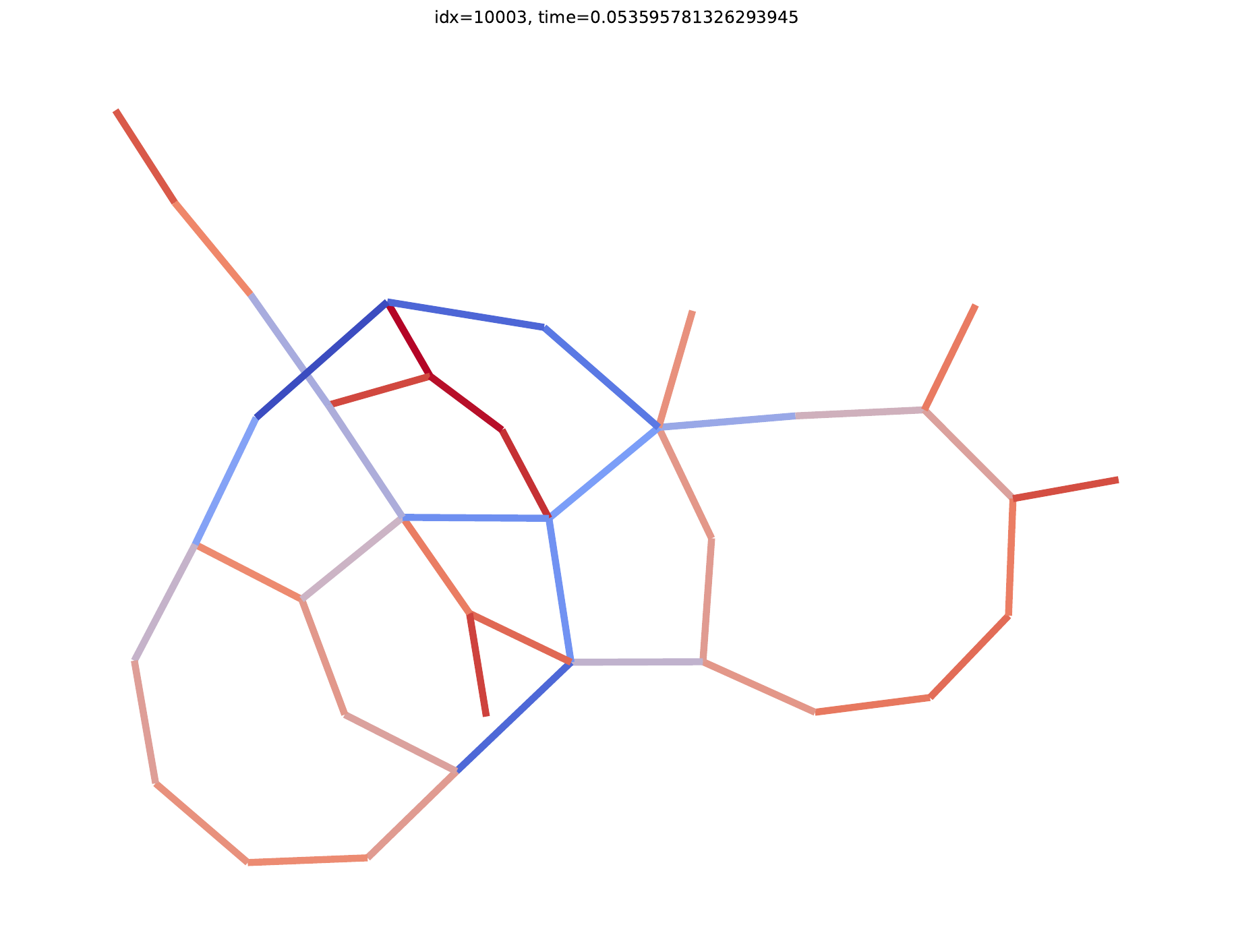} \\

&
t = 0.00s &
t = 0.33s &
t = 0.10s &
t = 0.05s &
t = 102.55s &
t = 0.03s &
t = 0.03s &
t = 0.06s &
t = 0.05s &
t = 0.03s &
t = 0.04s &
t = 0.05s \\

\makecell{\bfseries grafo9535.61\\N = 47\\M = 65} &
\imgcell{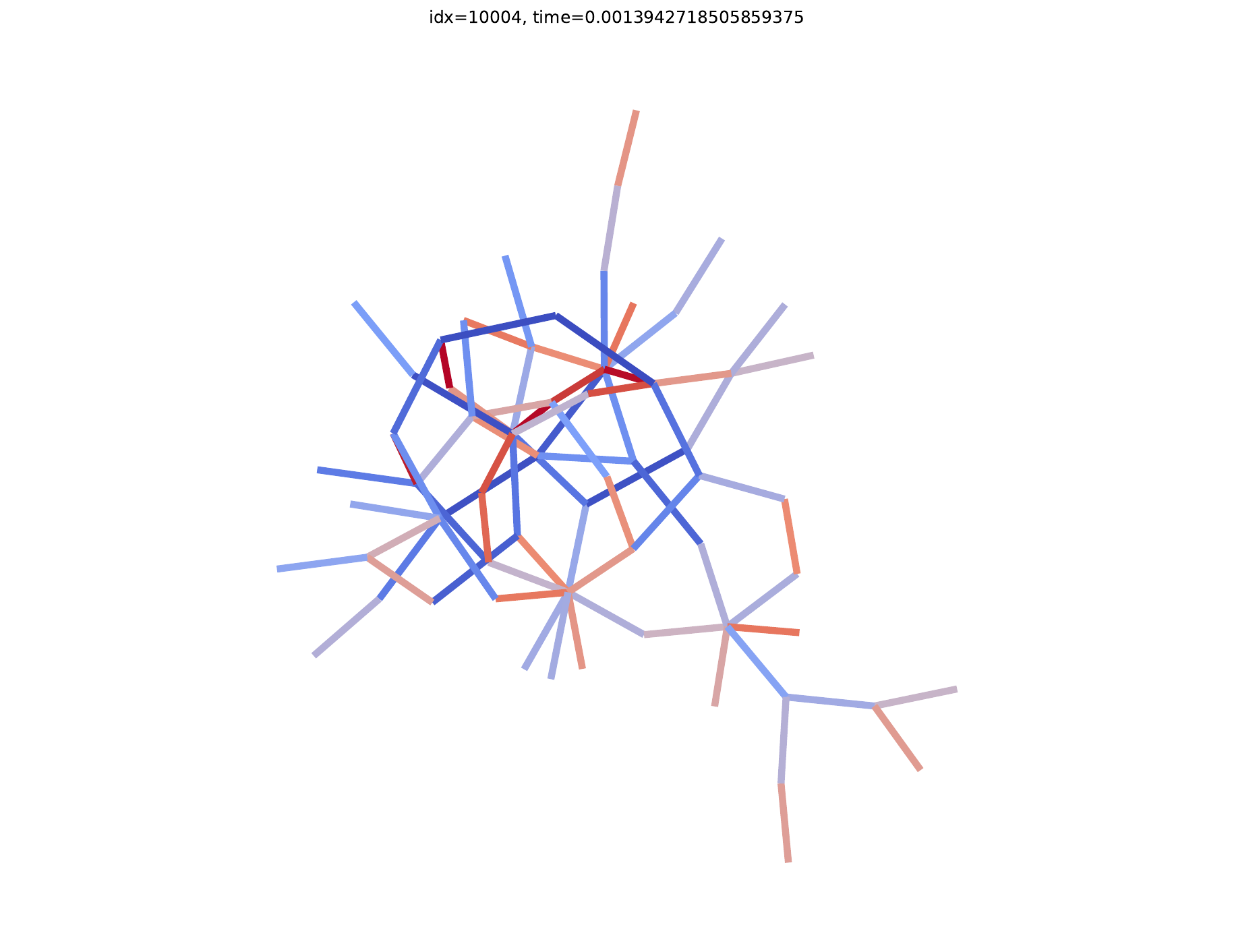} &
\imgcell{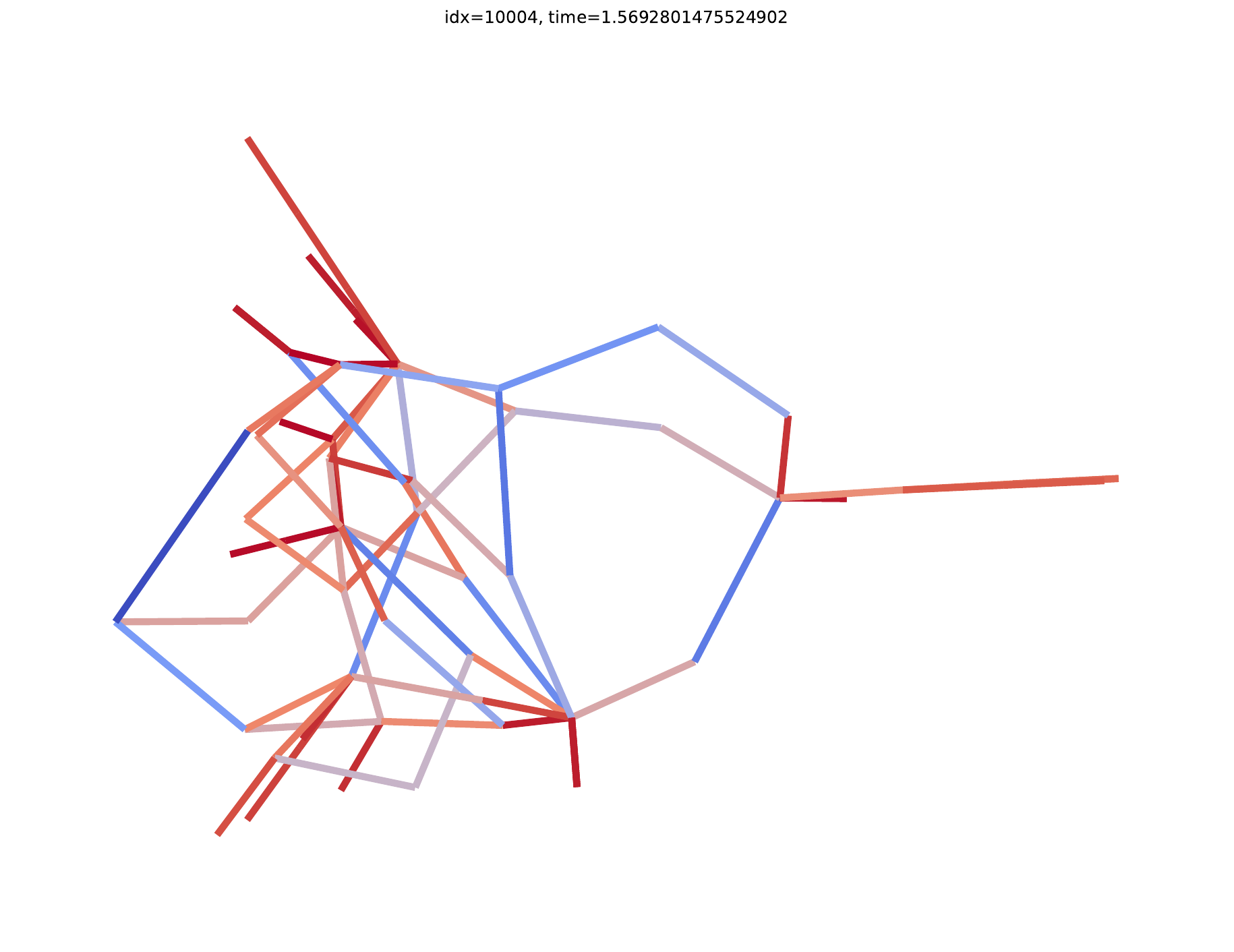} &
\imgcell{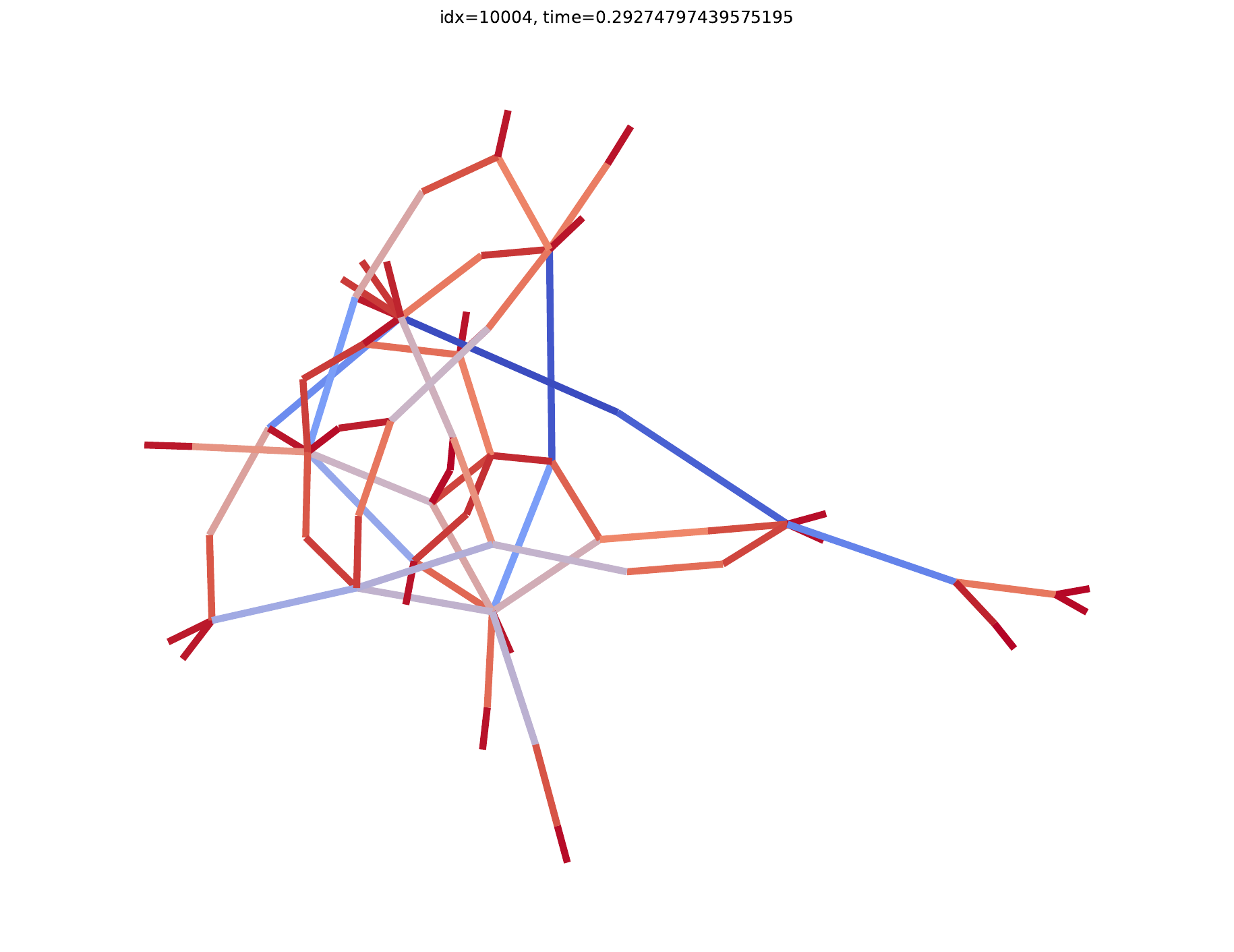} &
\imgcell{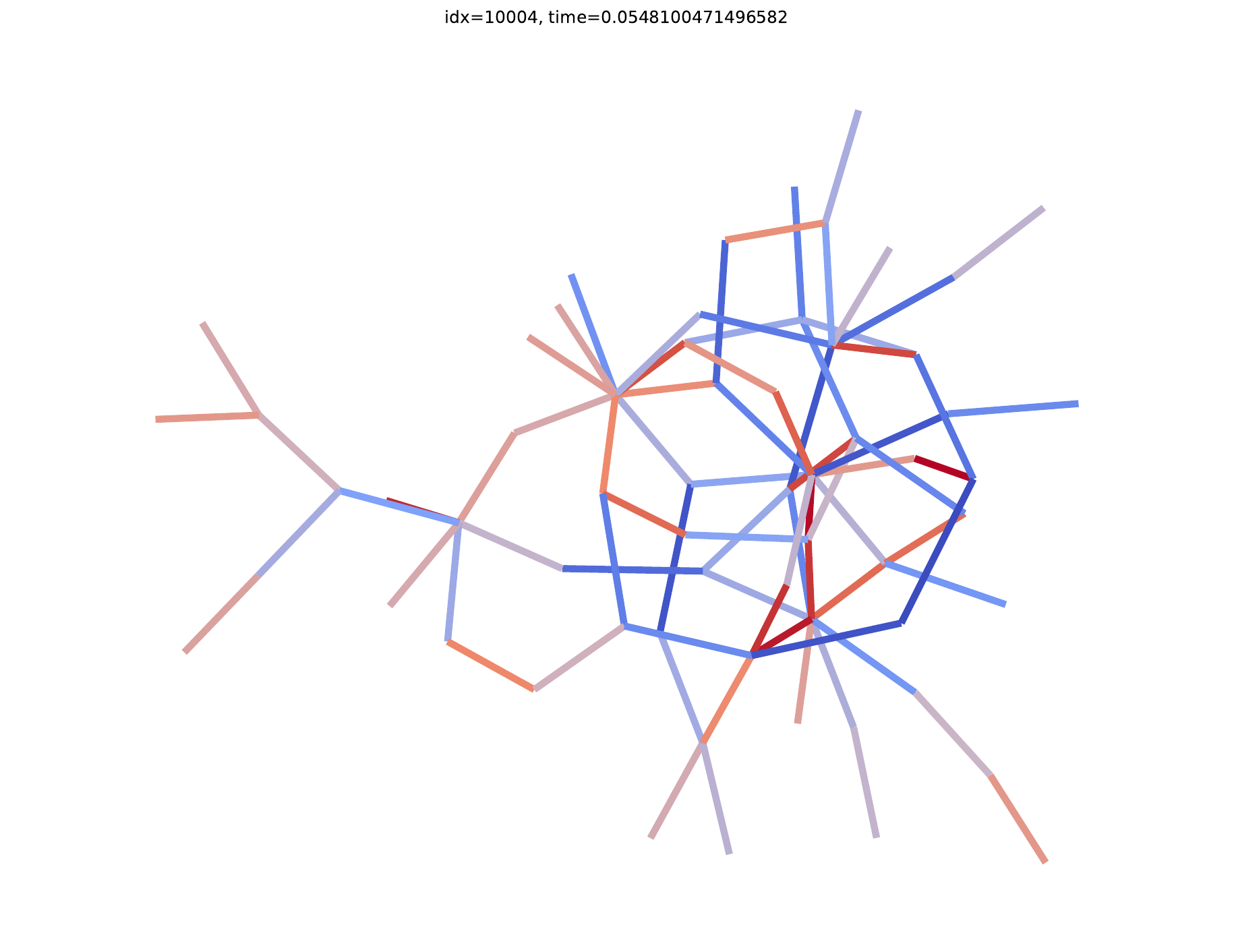} &
\imgcell{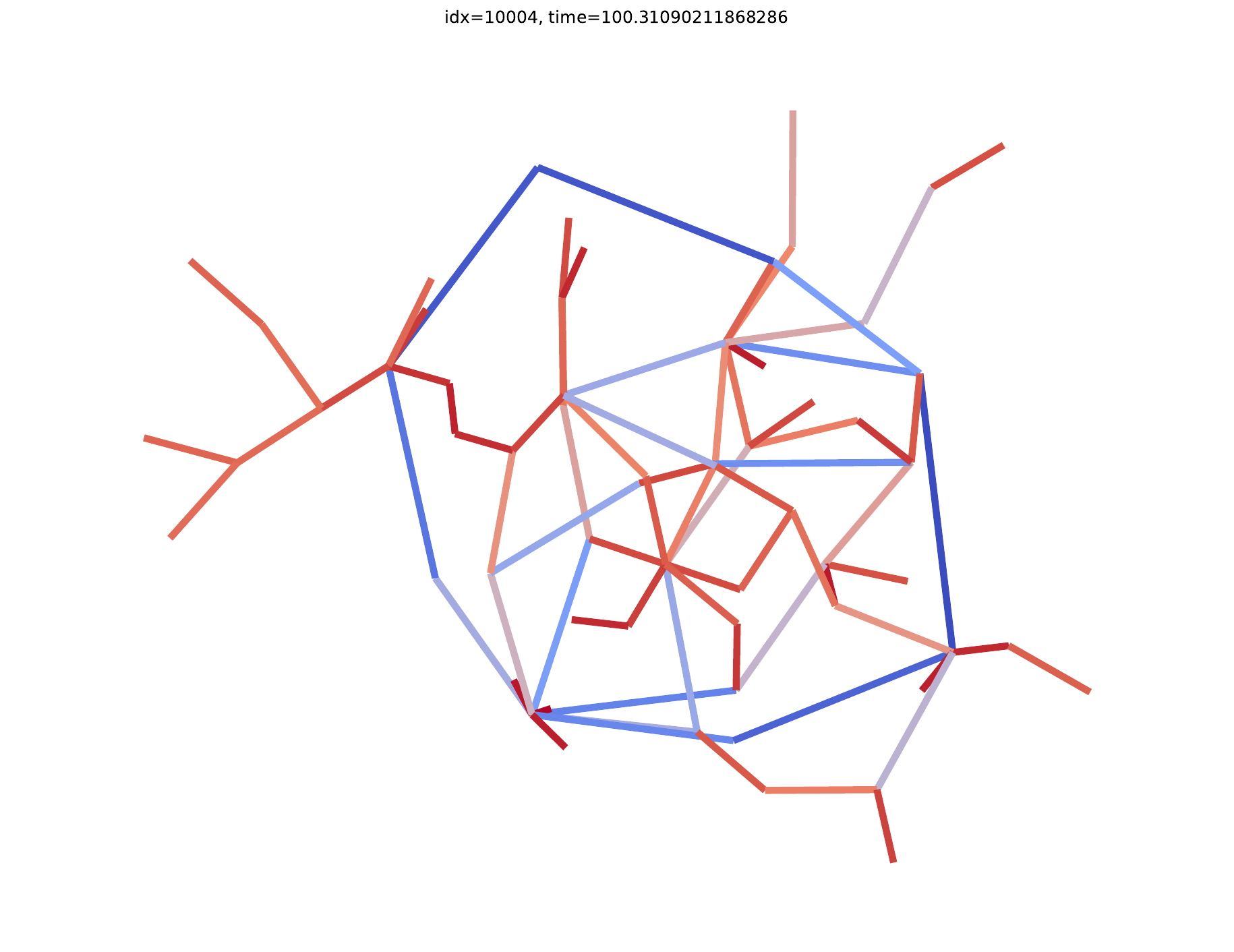} &
\imgcell{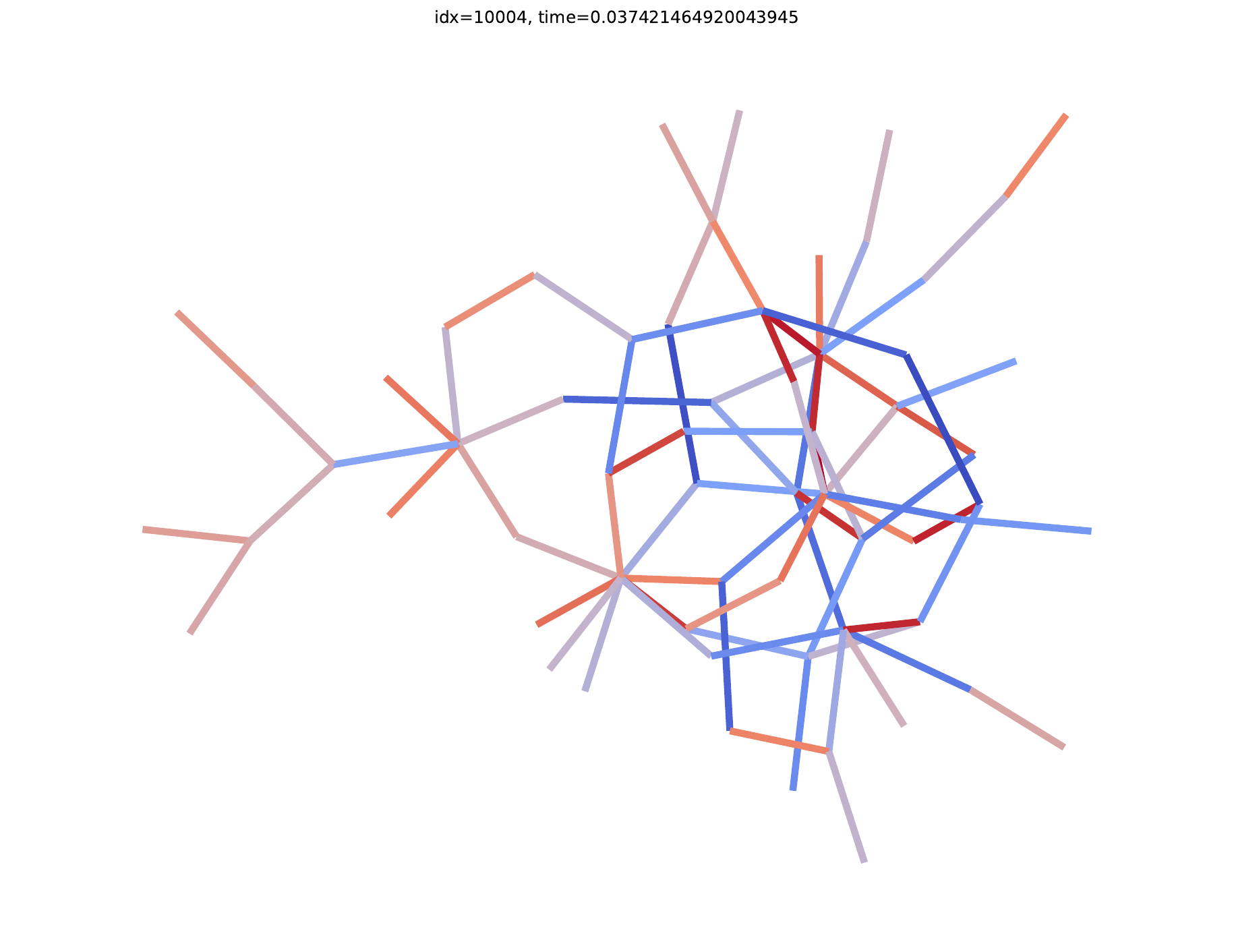} &
\imgcell{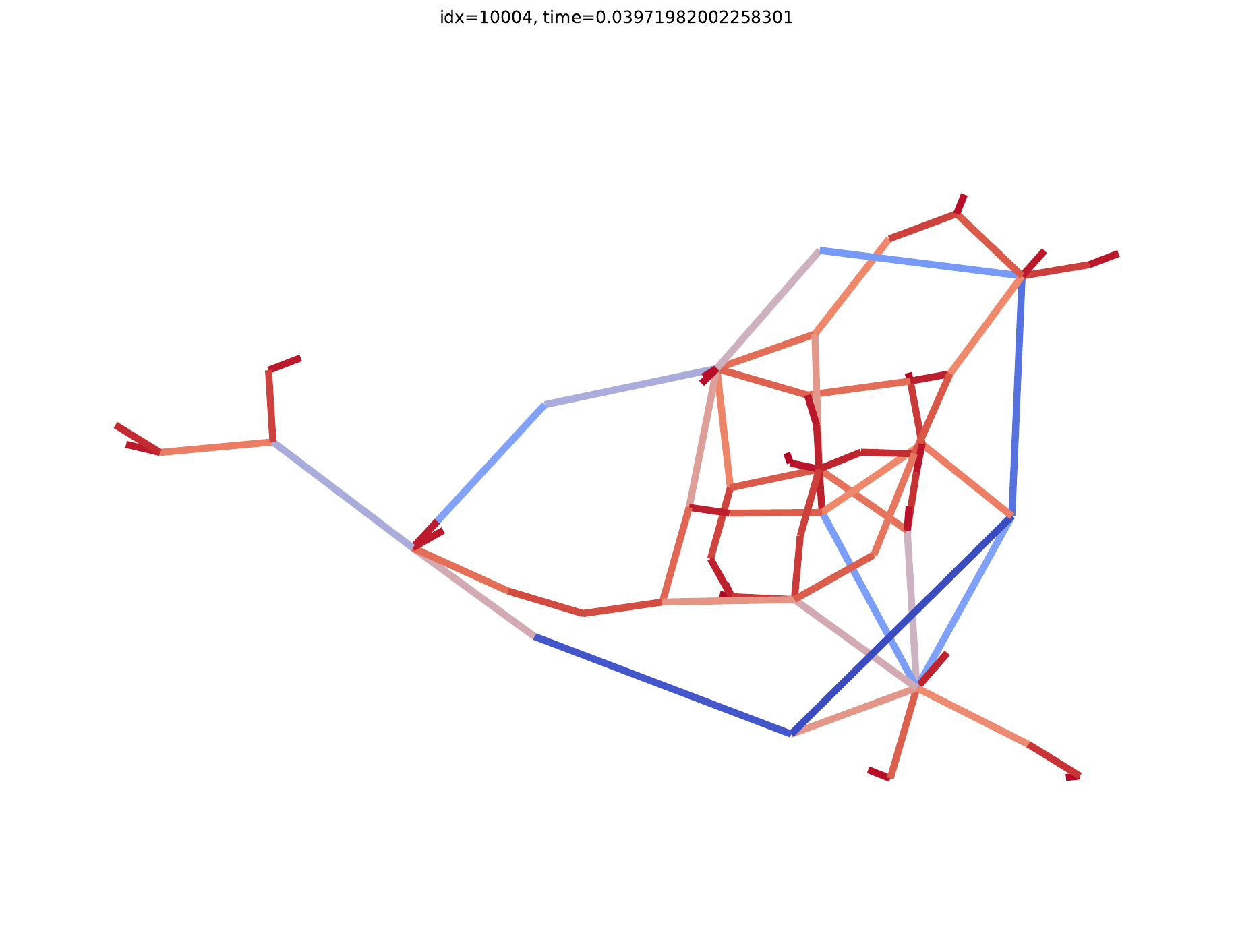} &
\imgcell{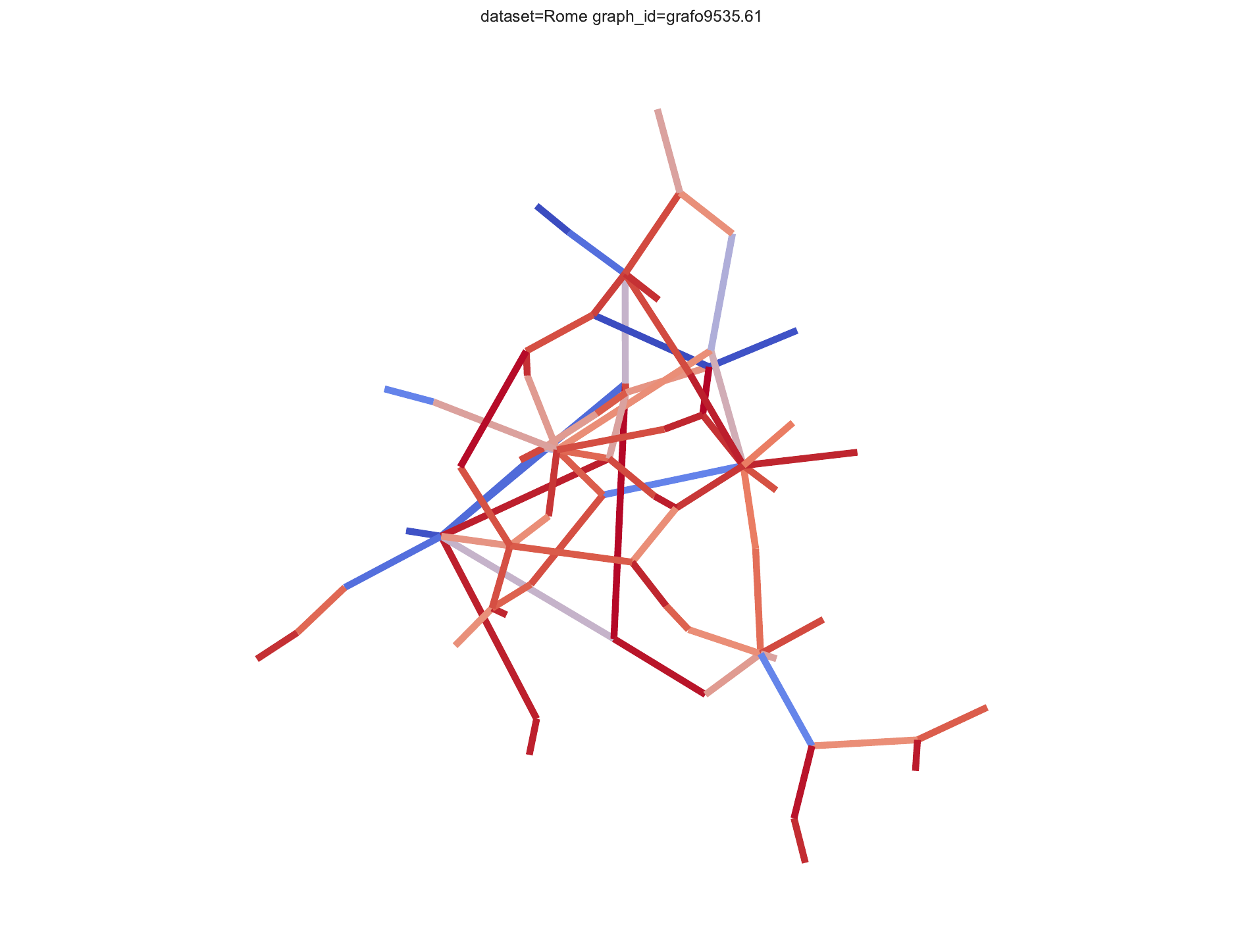} &
\imgcell{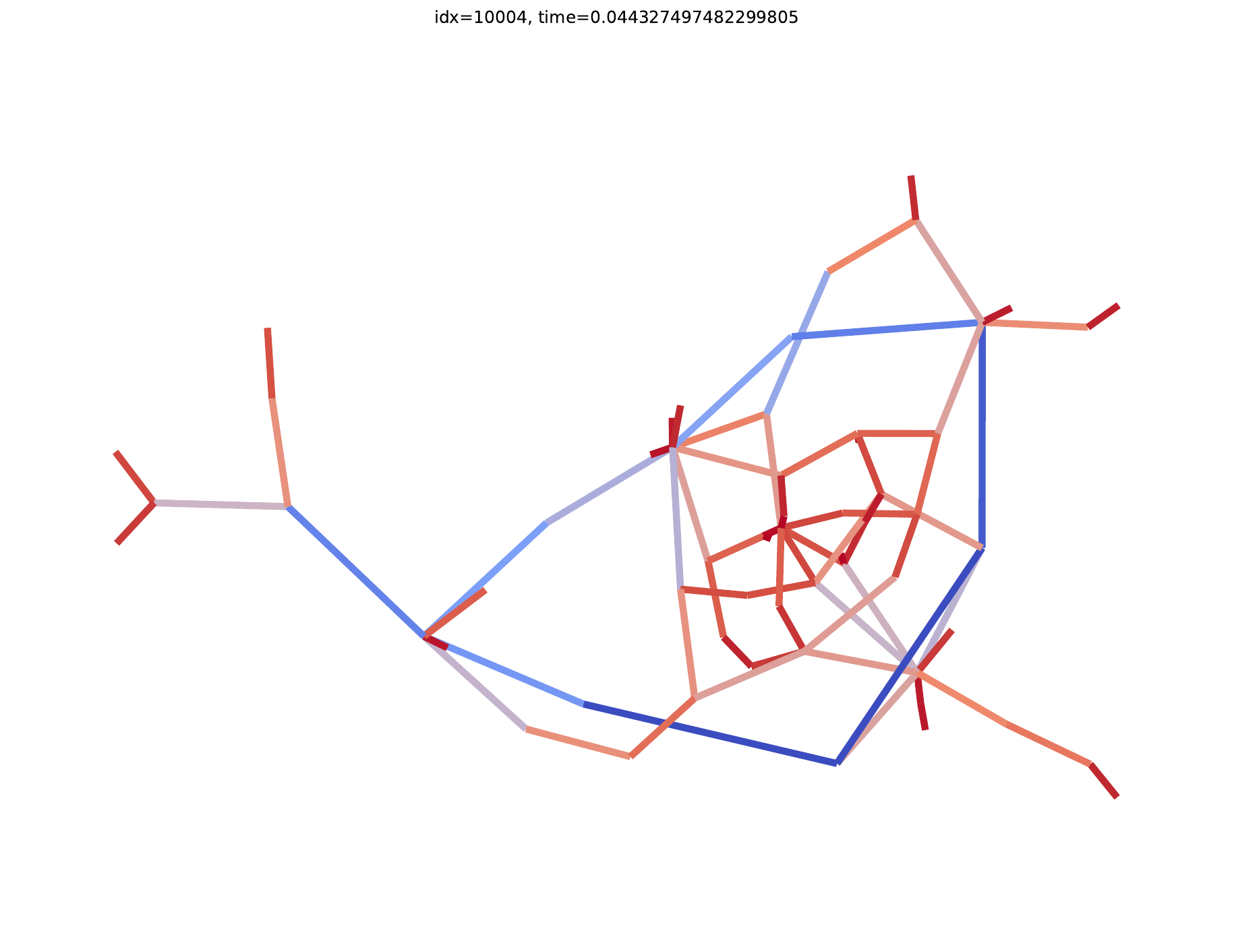} &
\imgcell{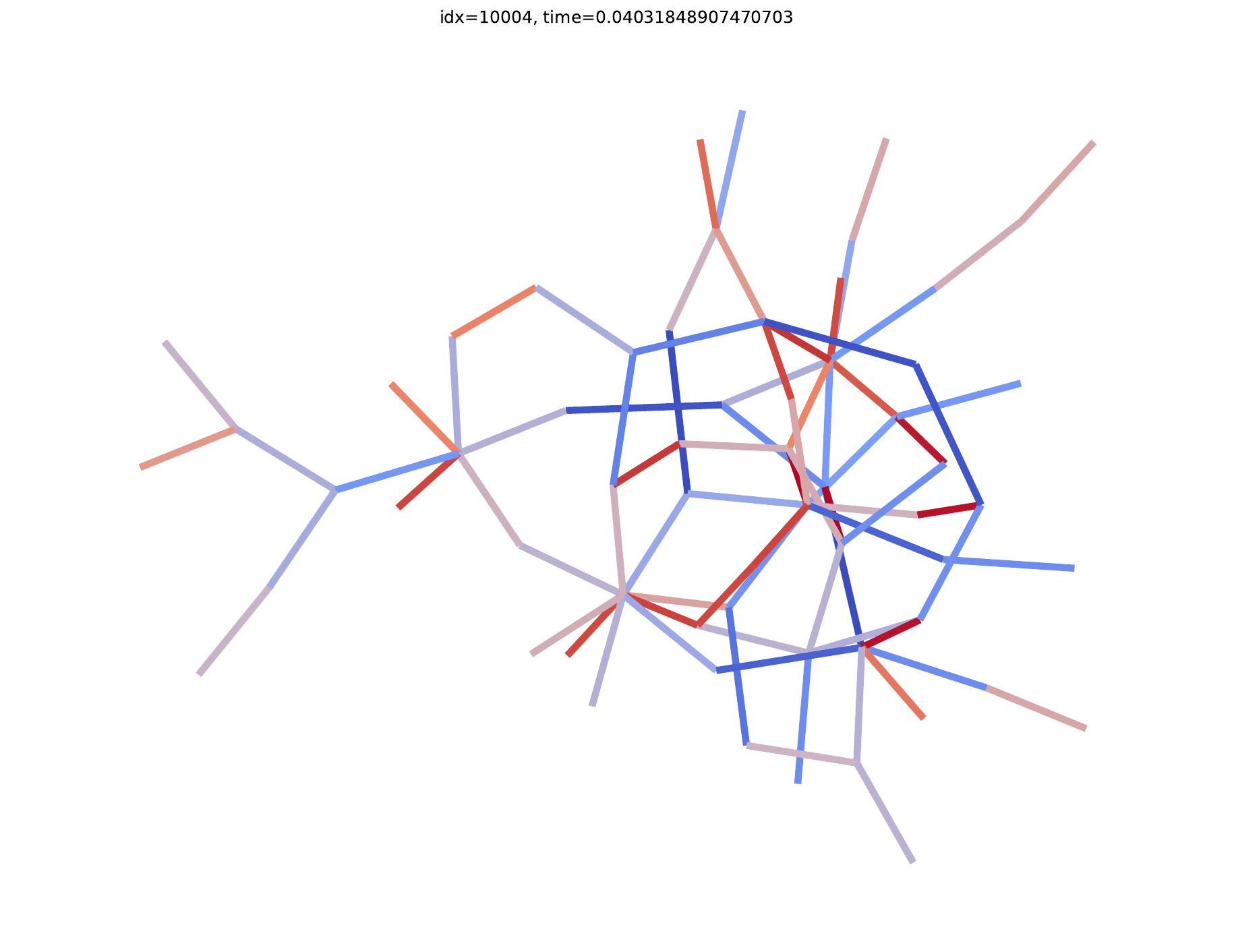} &
\imgcell{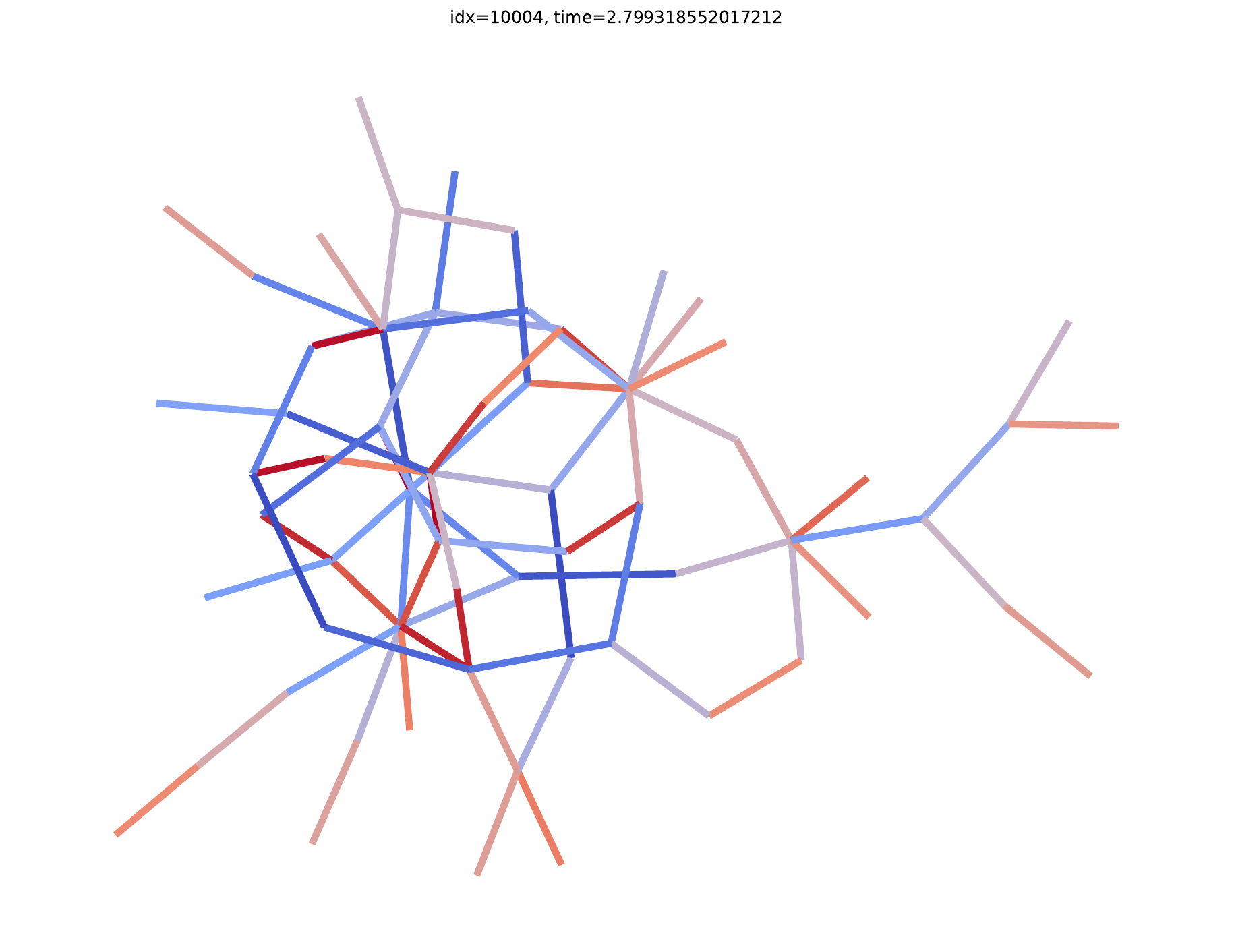} &
\imgcell{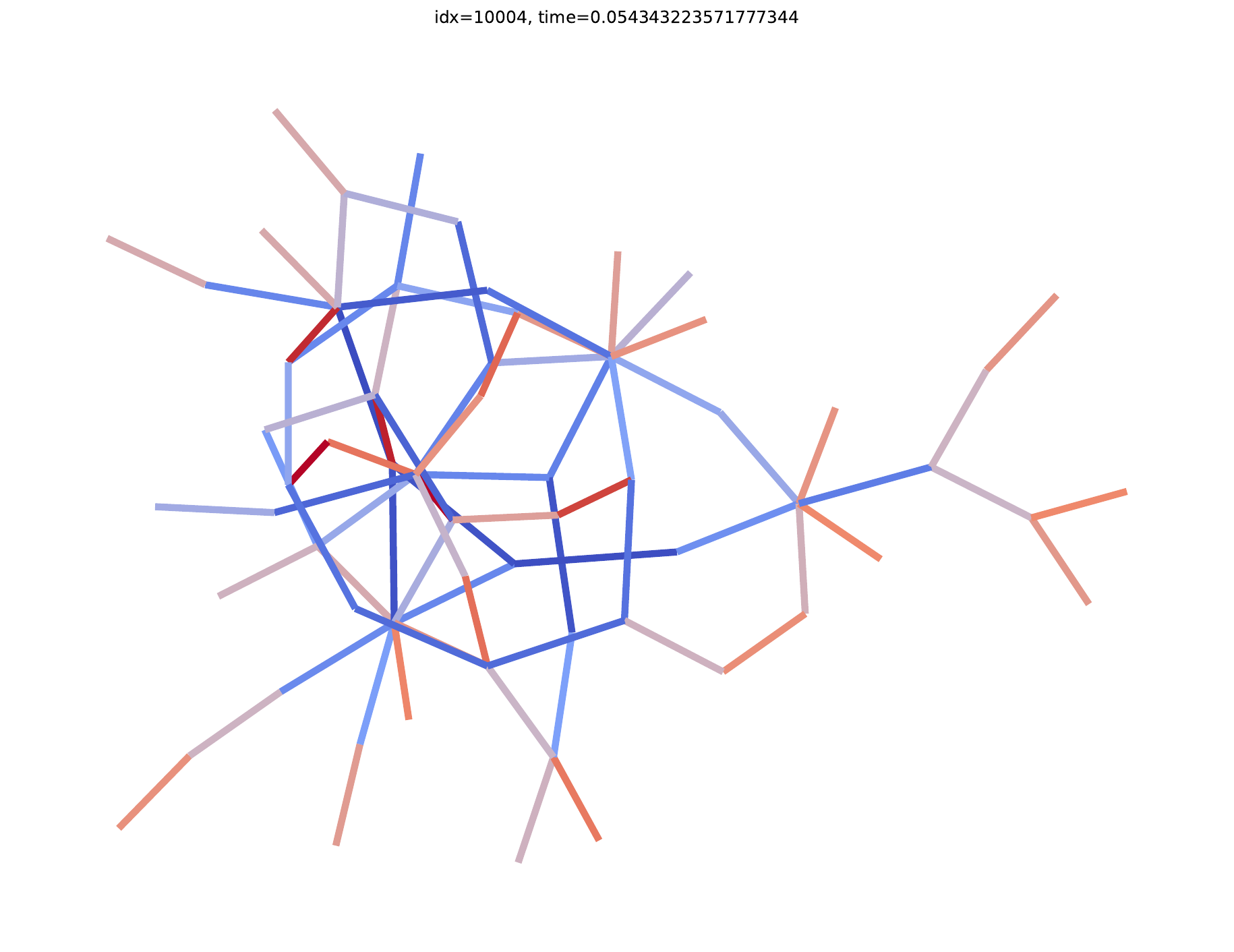} \\

&
t = 0.00s &
t = 1.57s &
t = 0.29s &
t = 0.05s &
t = 100.31s &
t = 0.04s &
t = 0.04s &
t = 0.04s &
t = 0.04s &
t = 0.04s &
t = 0.03s &
t = 0.05s \\

\makecell{\bfseries grafo7796.93\\N = 37\\M = 46} &
\imgcell{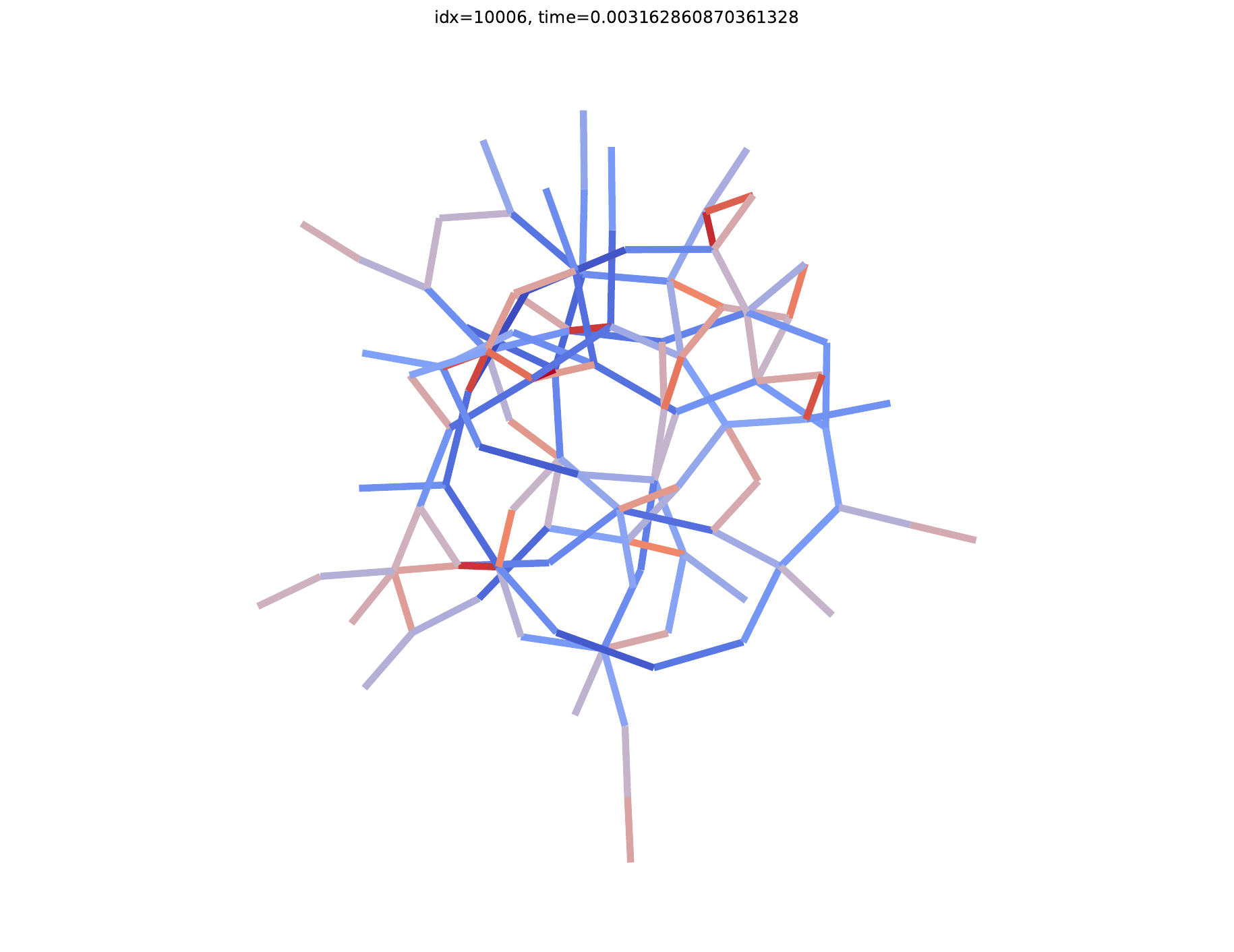} &
\imgcell{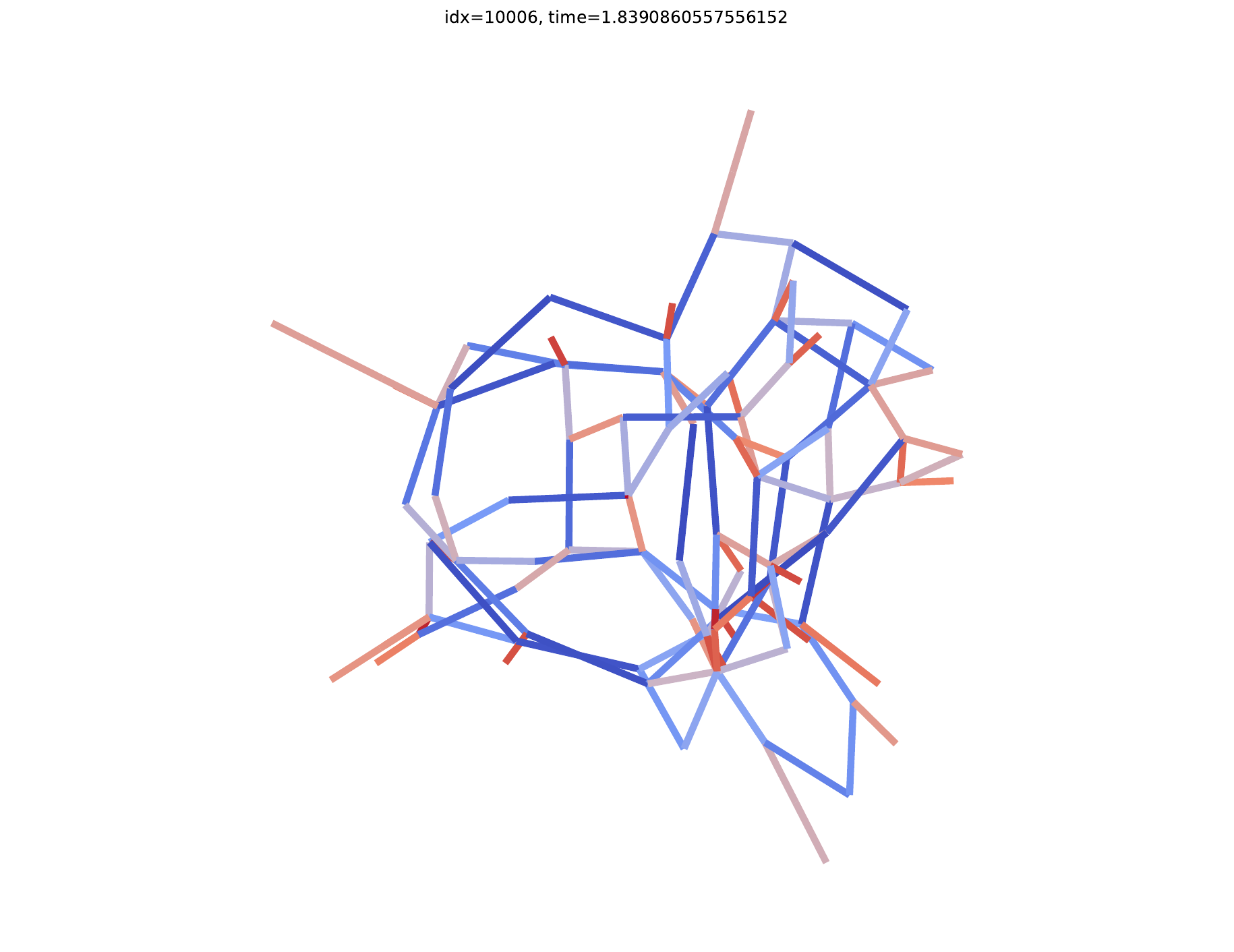} &
\imgcell{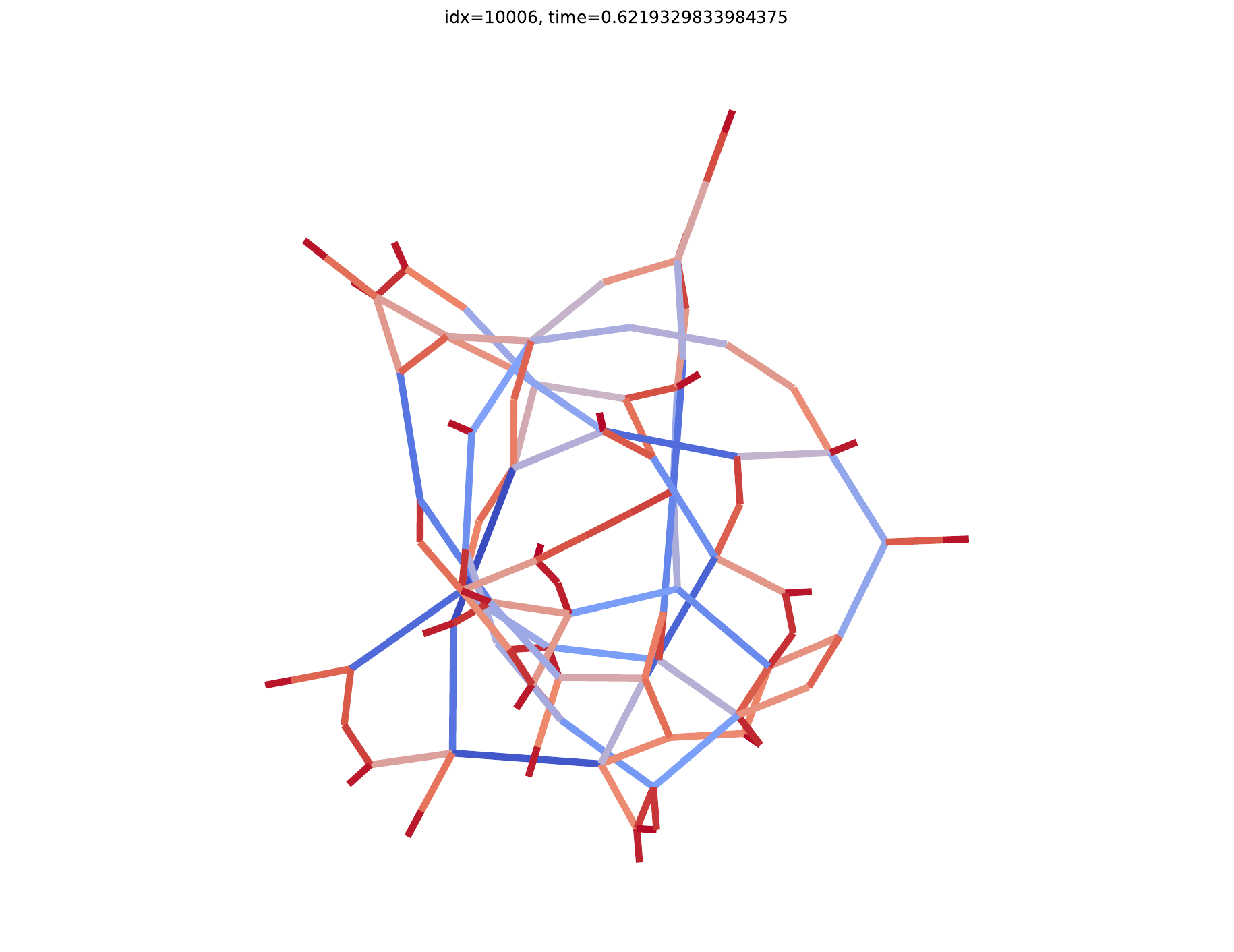} &
\imgcell{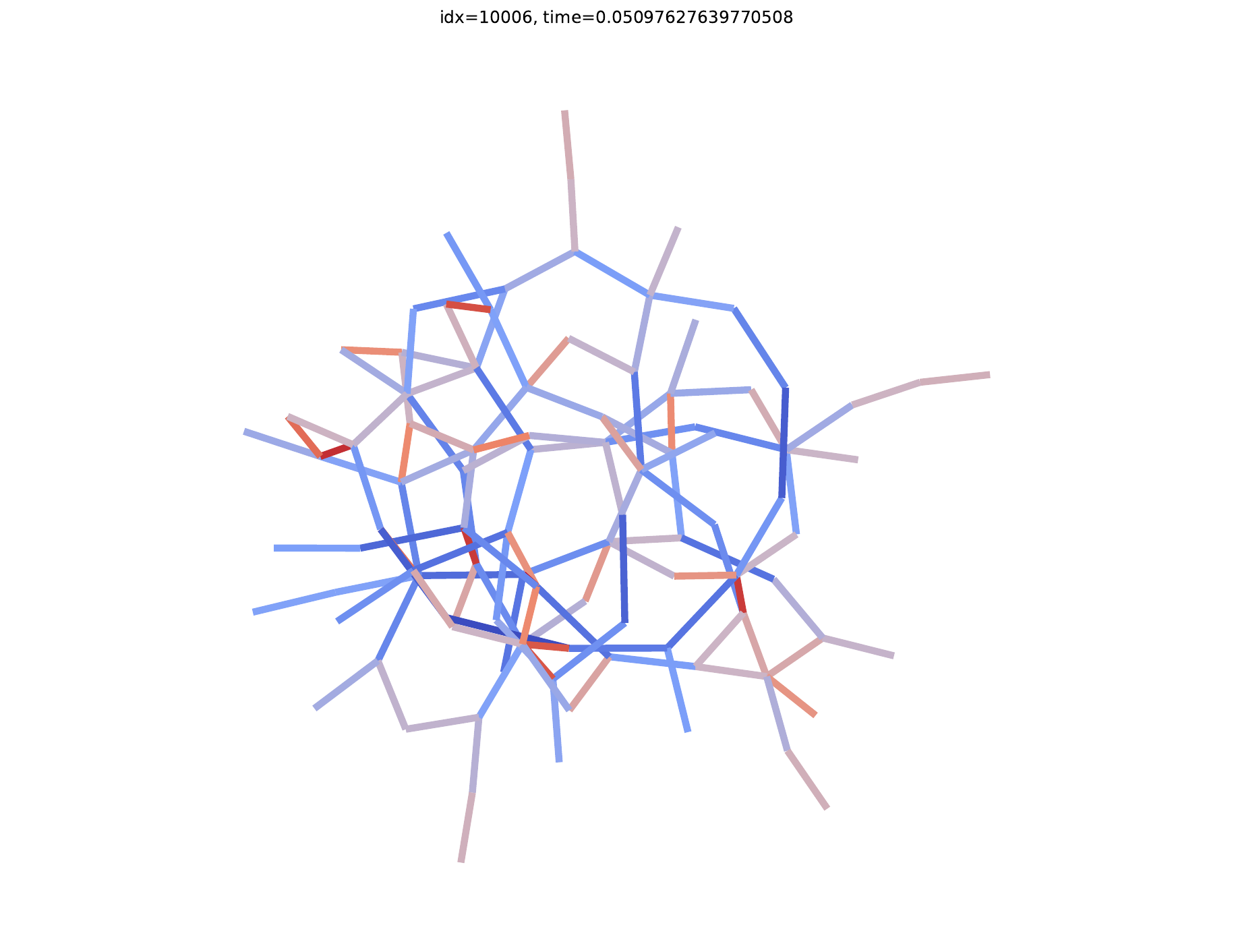} &
\imgcell{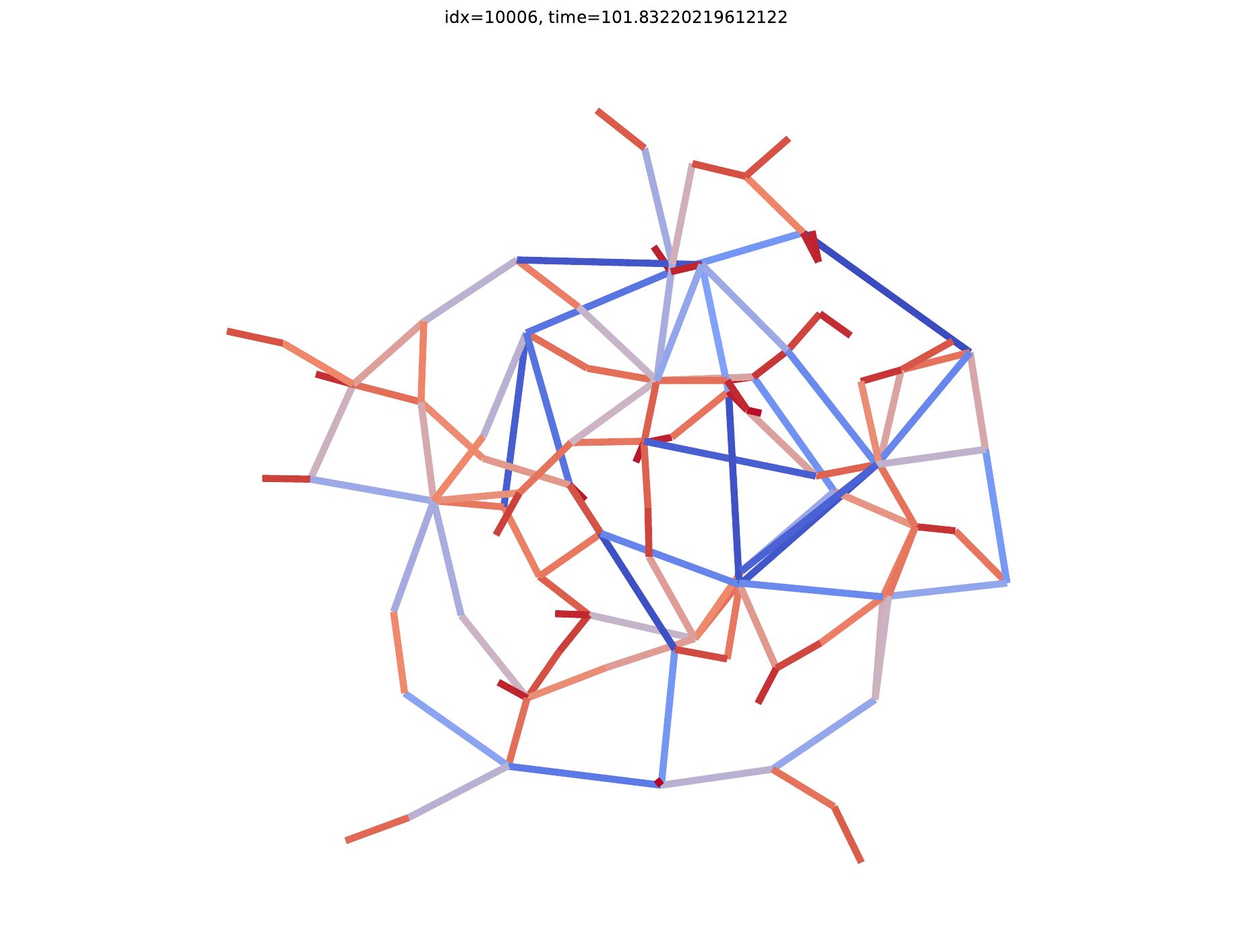} &
\imgcell{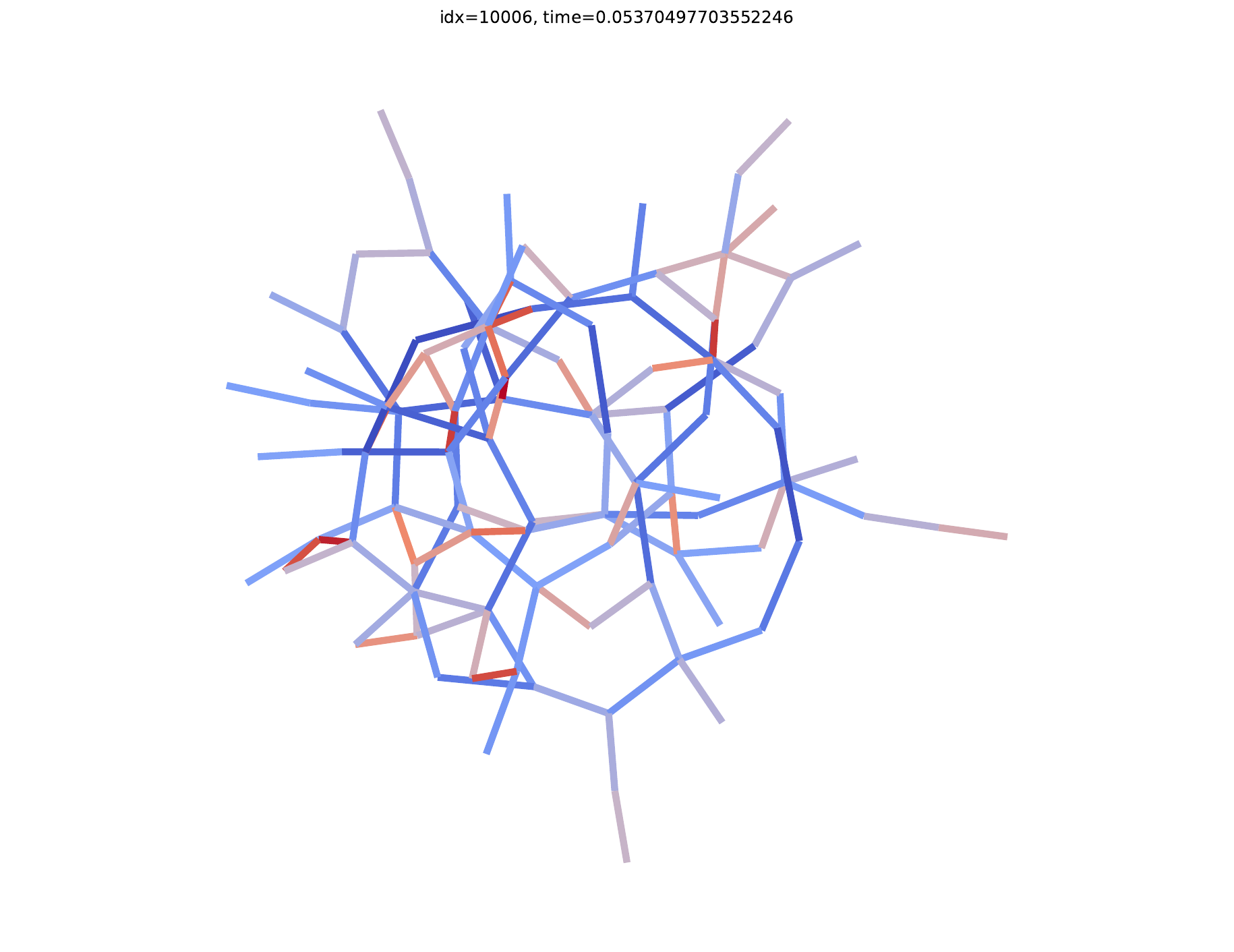} &
\imgcell{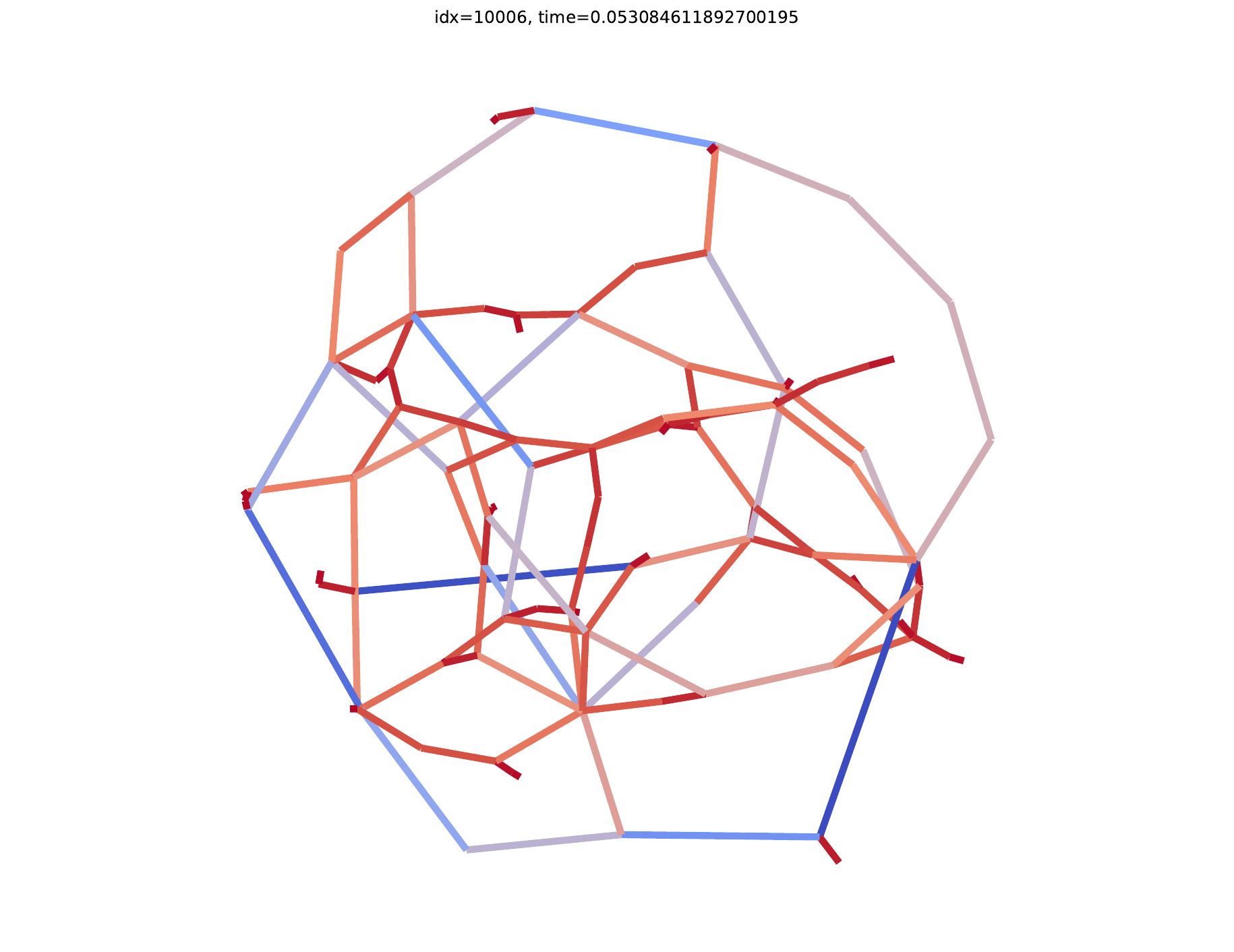} &
\imgcell{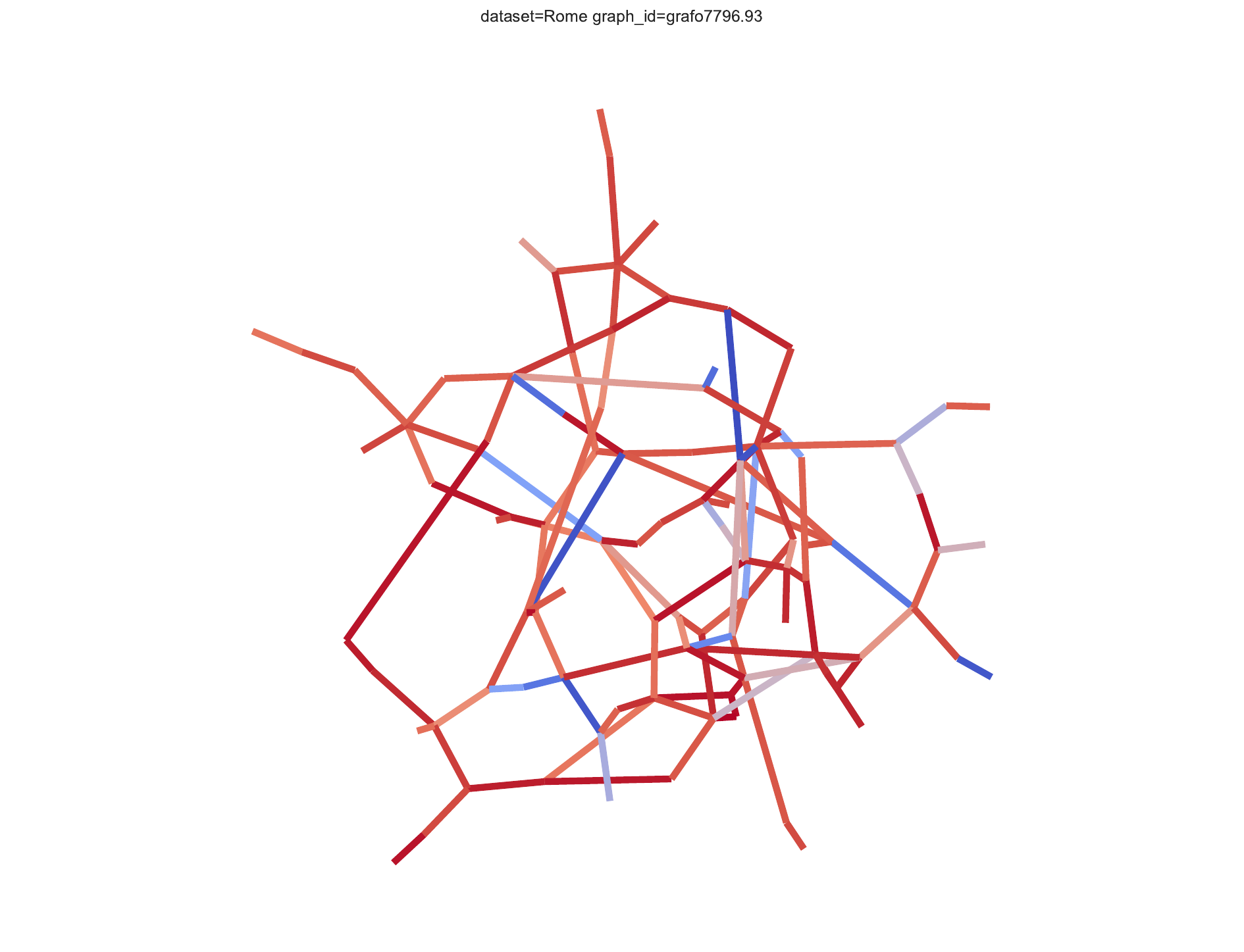} &
\imgcell{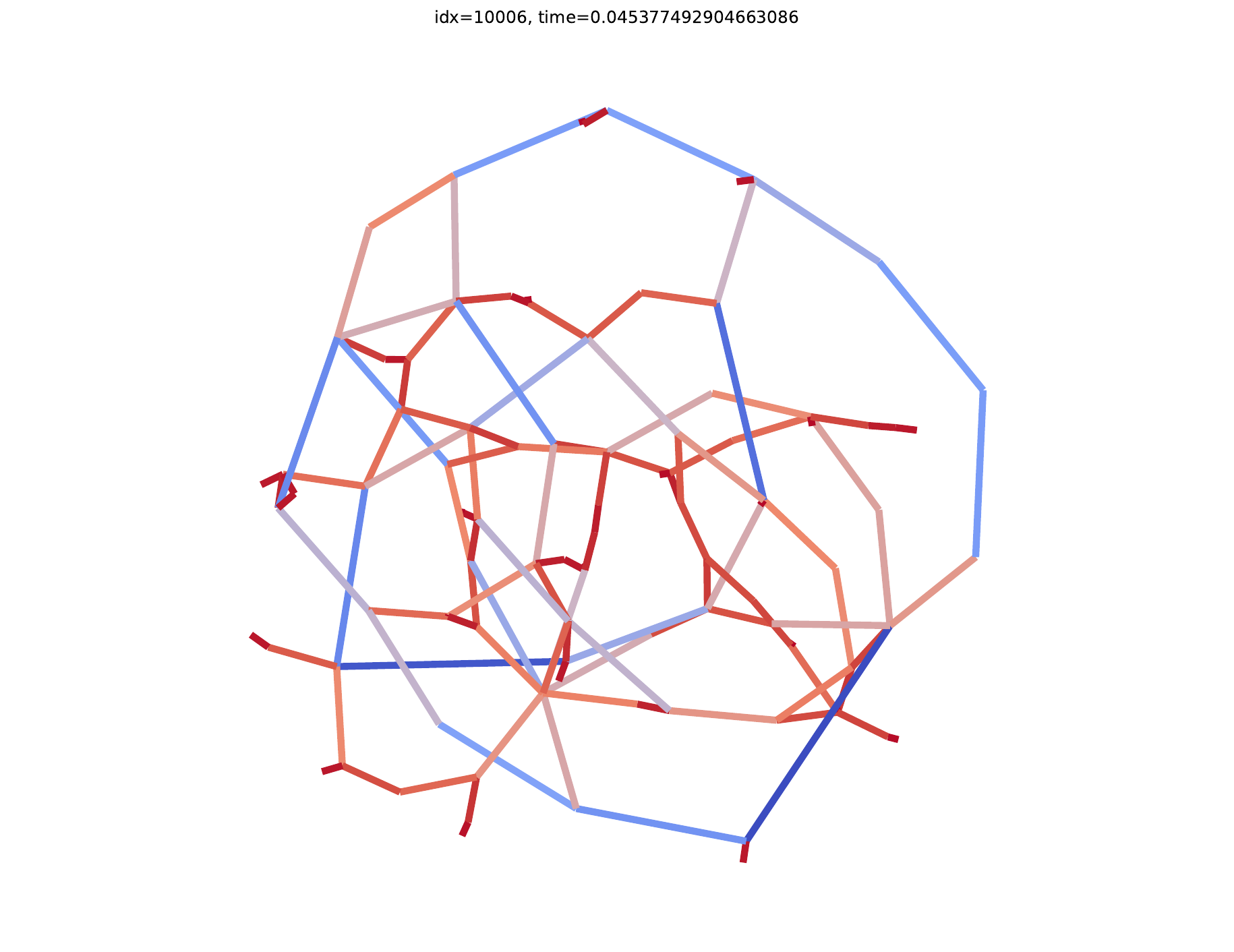} &
\imgcell{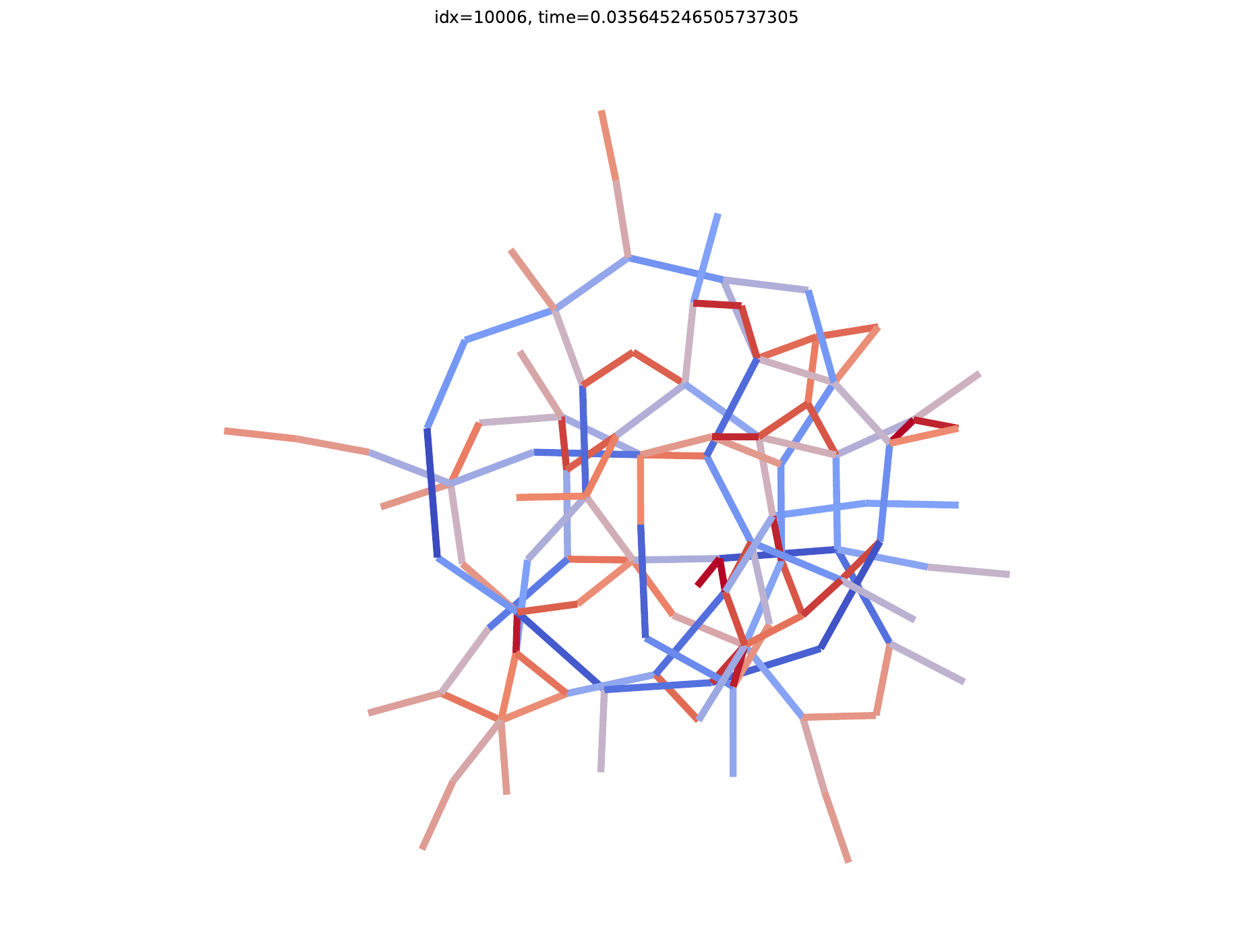} &
\imgcell{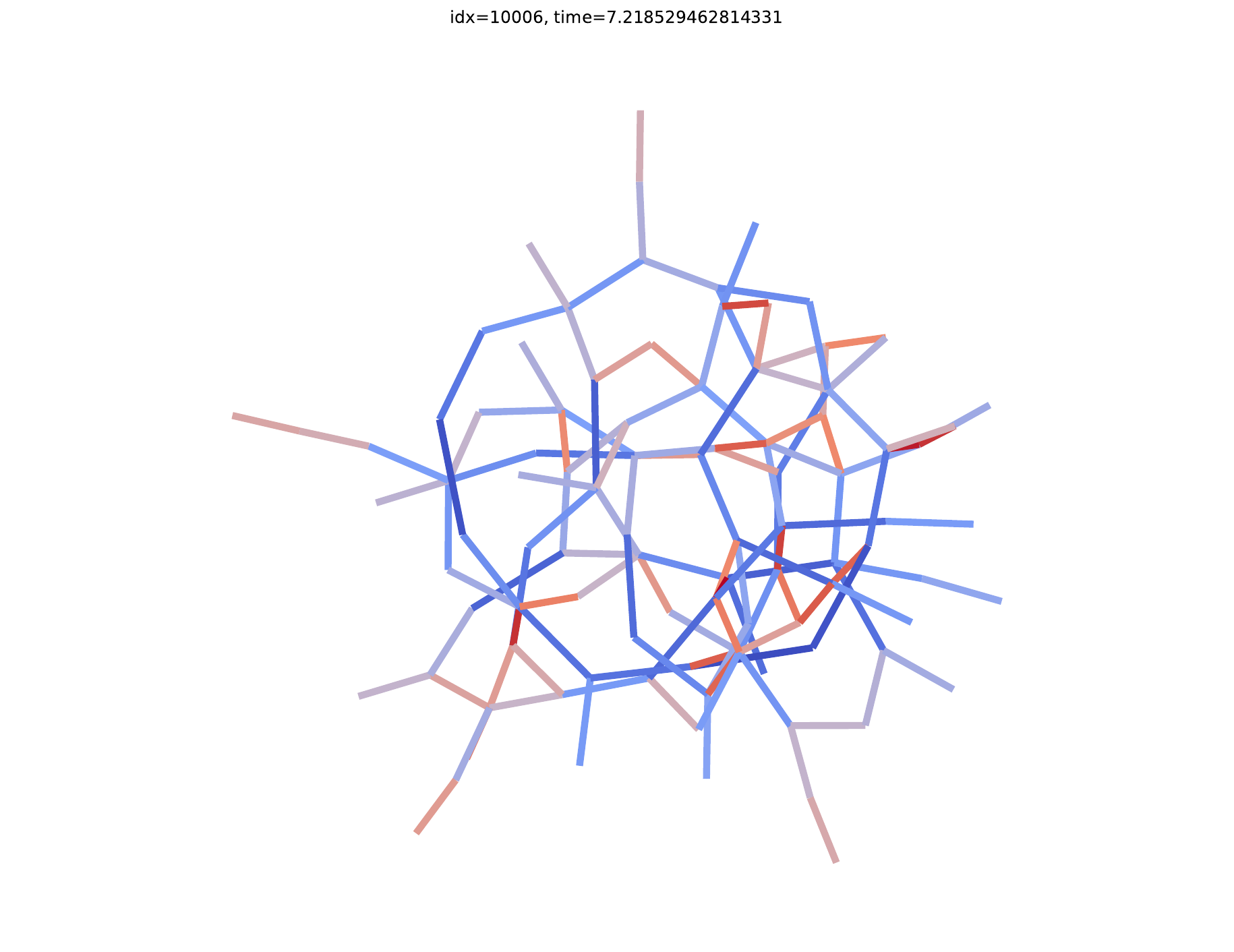} &
\imgcell{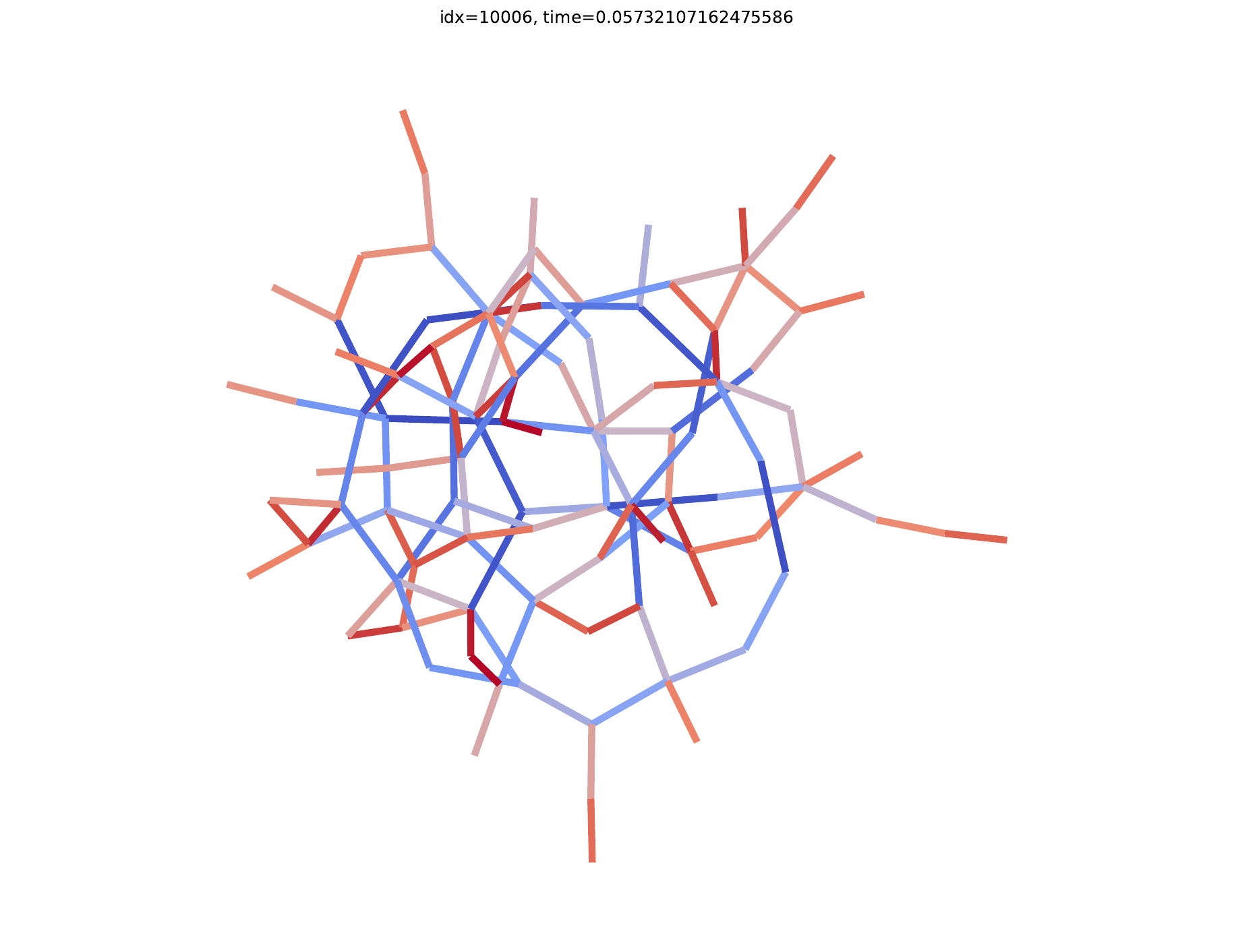} \\

&
t = 0.00s &
t = 1.84s &
t = 0.62s &
t = 0.05s &
t = 101.83s &
t = 0.05s &
t = 0.05s &
t = 0.04s &
t = 0.05s &
t = 0.04s &
t = 0.05s &
t = 0.06s \\

\makecell{\bfseries grafo10131.98\\N = 39\\M = 49} &
\imgcell{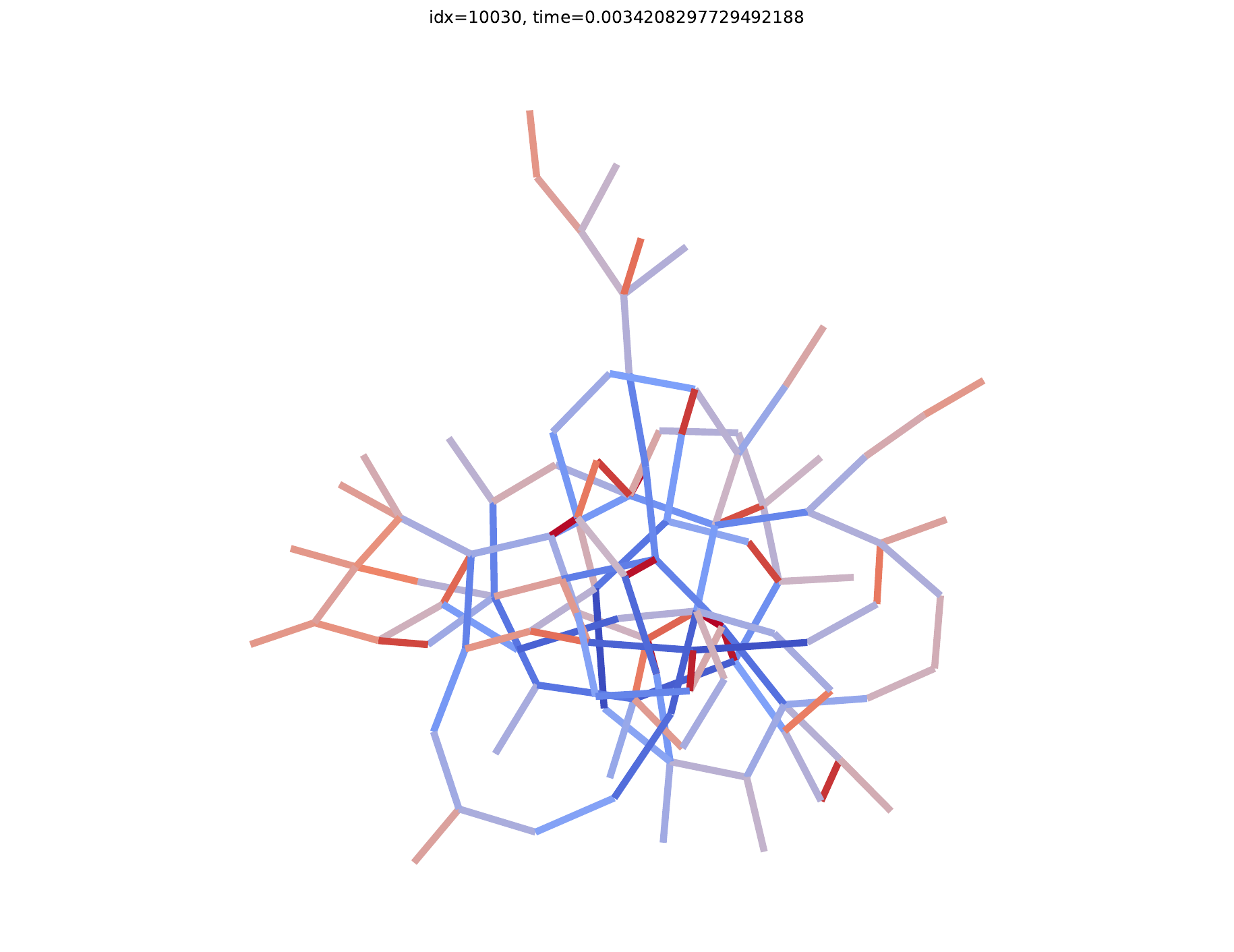} &
\imgcell{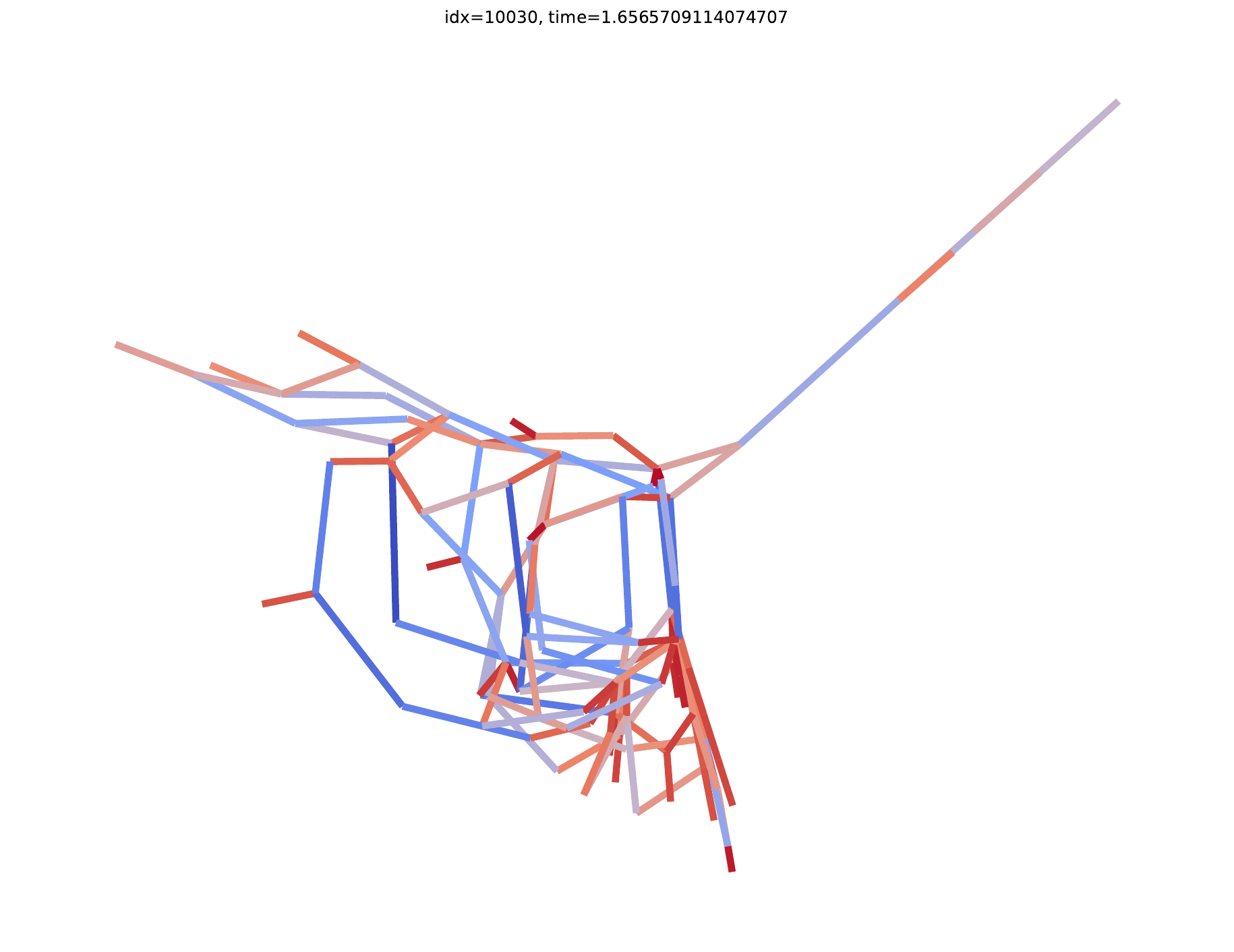} &
\imgcell{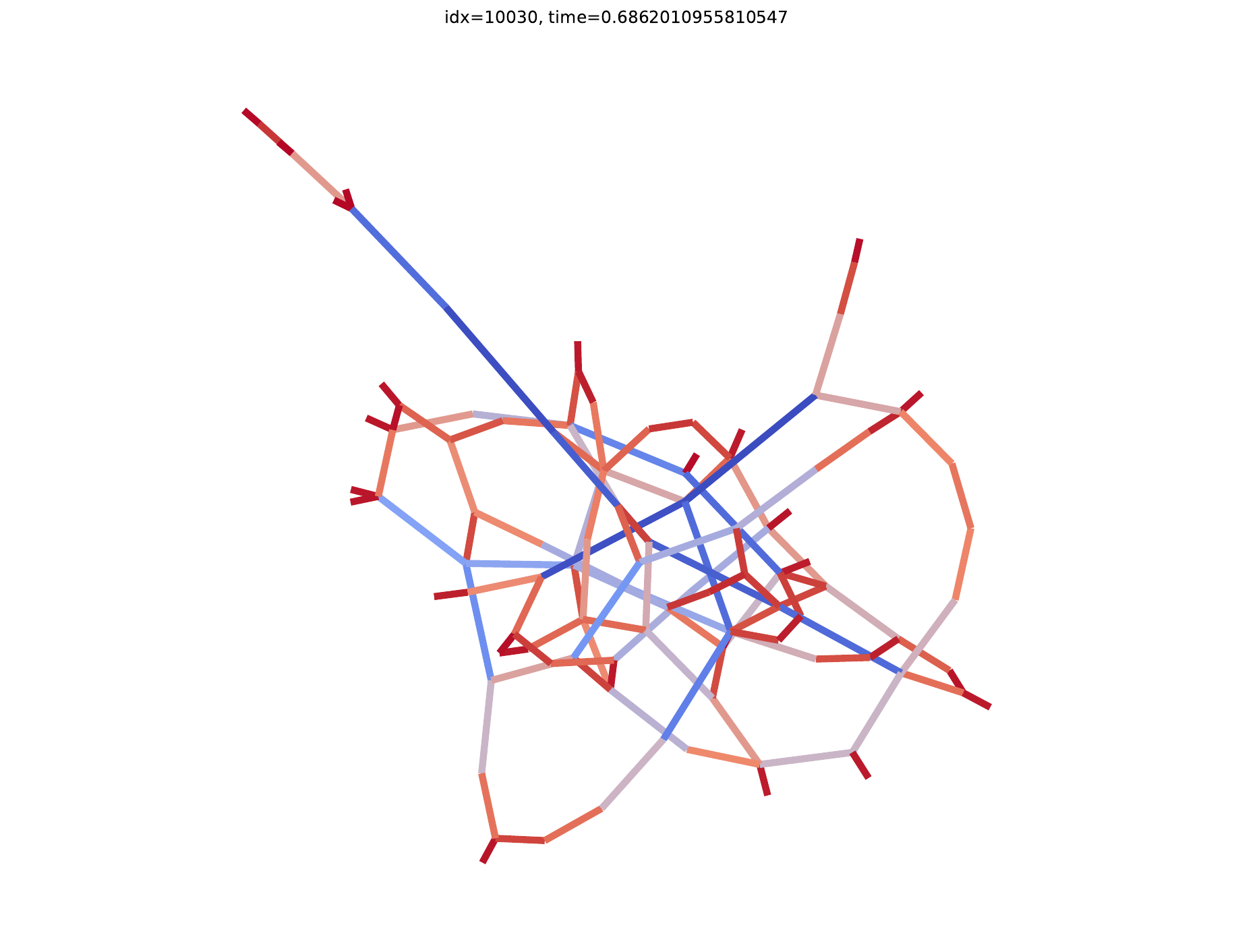} &
\imgcell{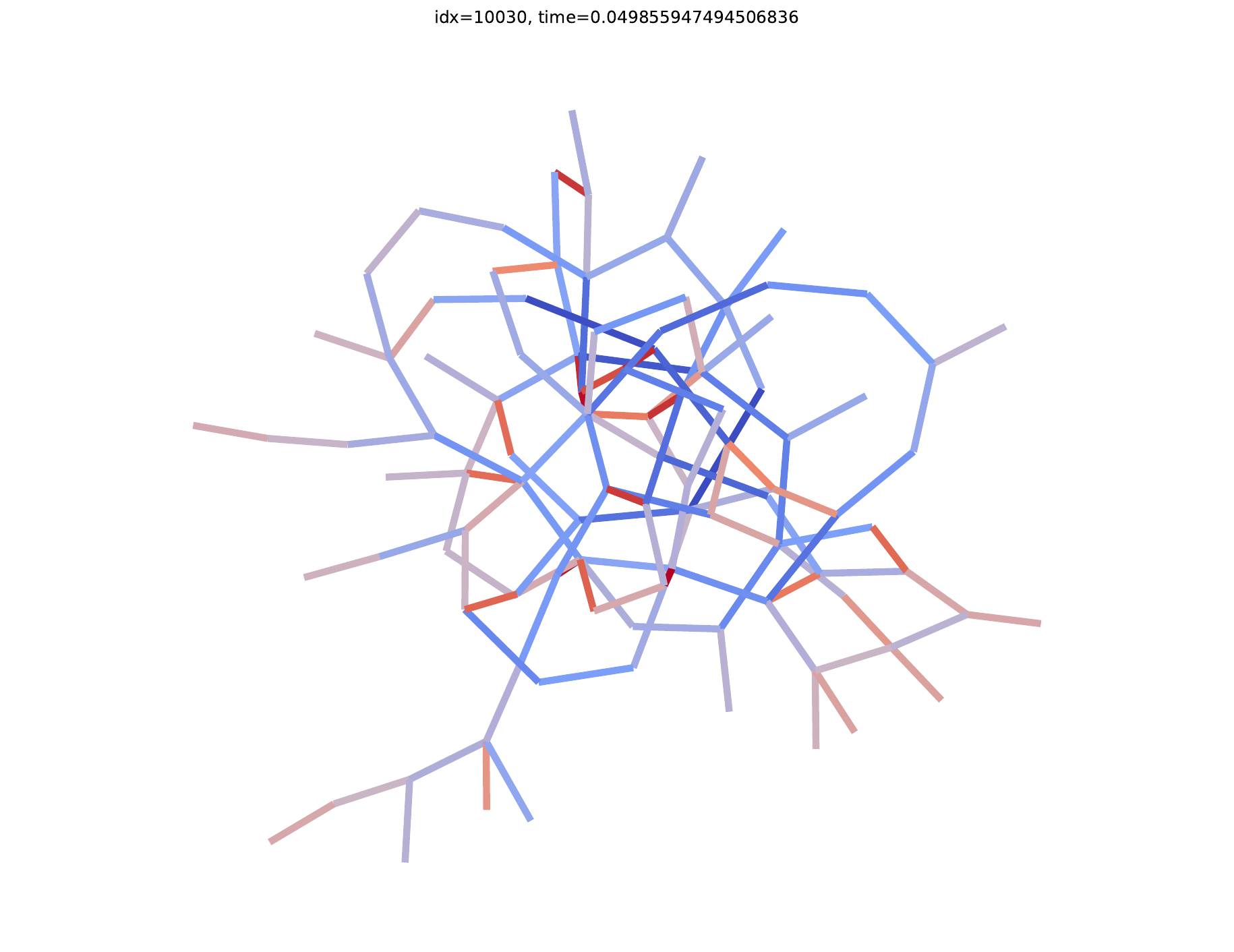} &
\imgcell{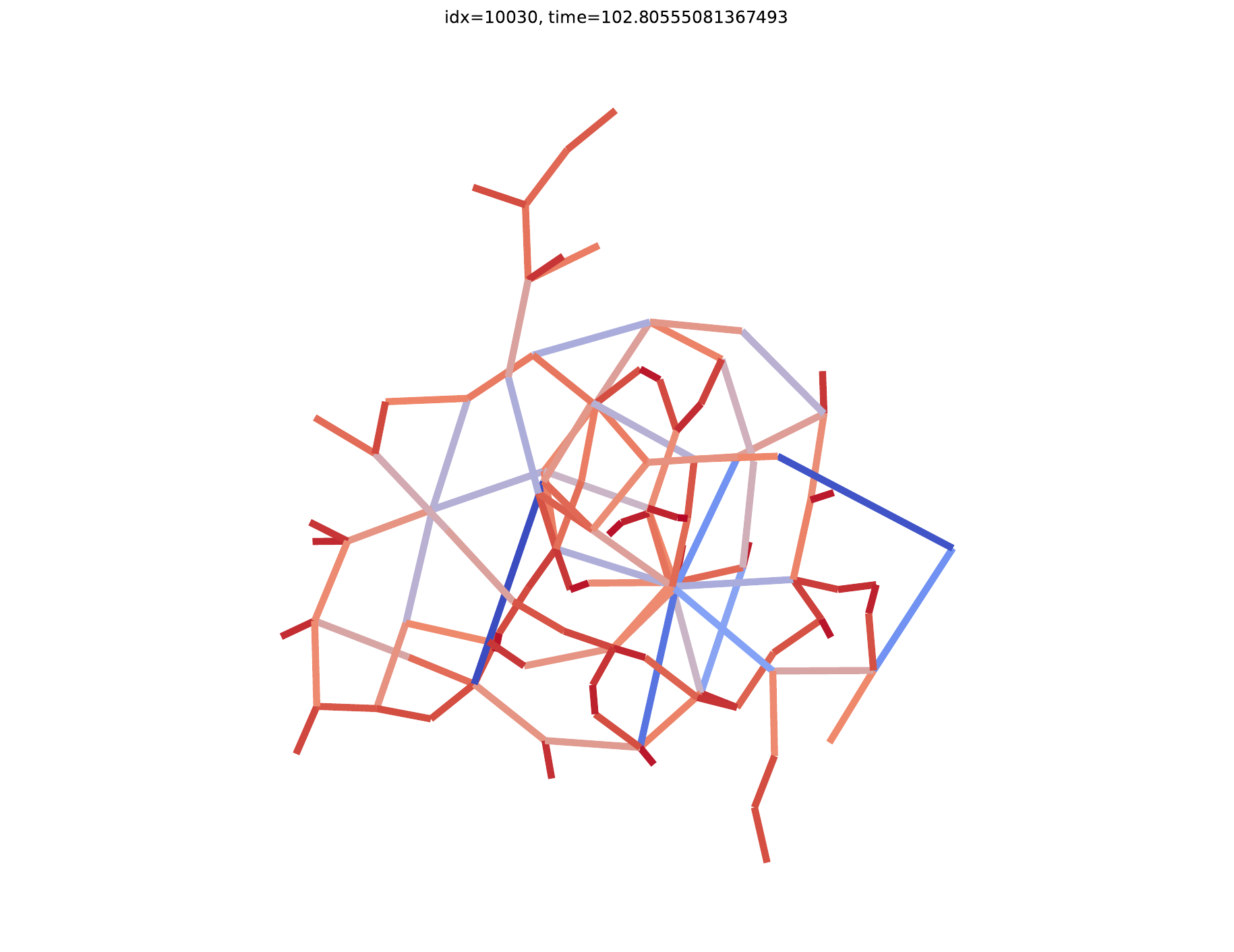} &
\imgcell{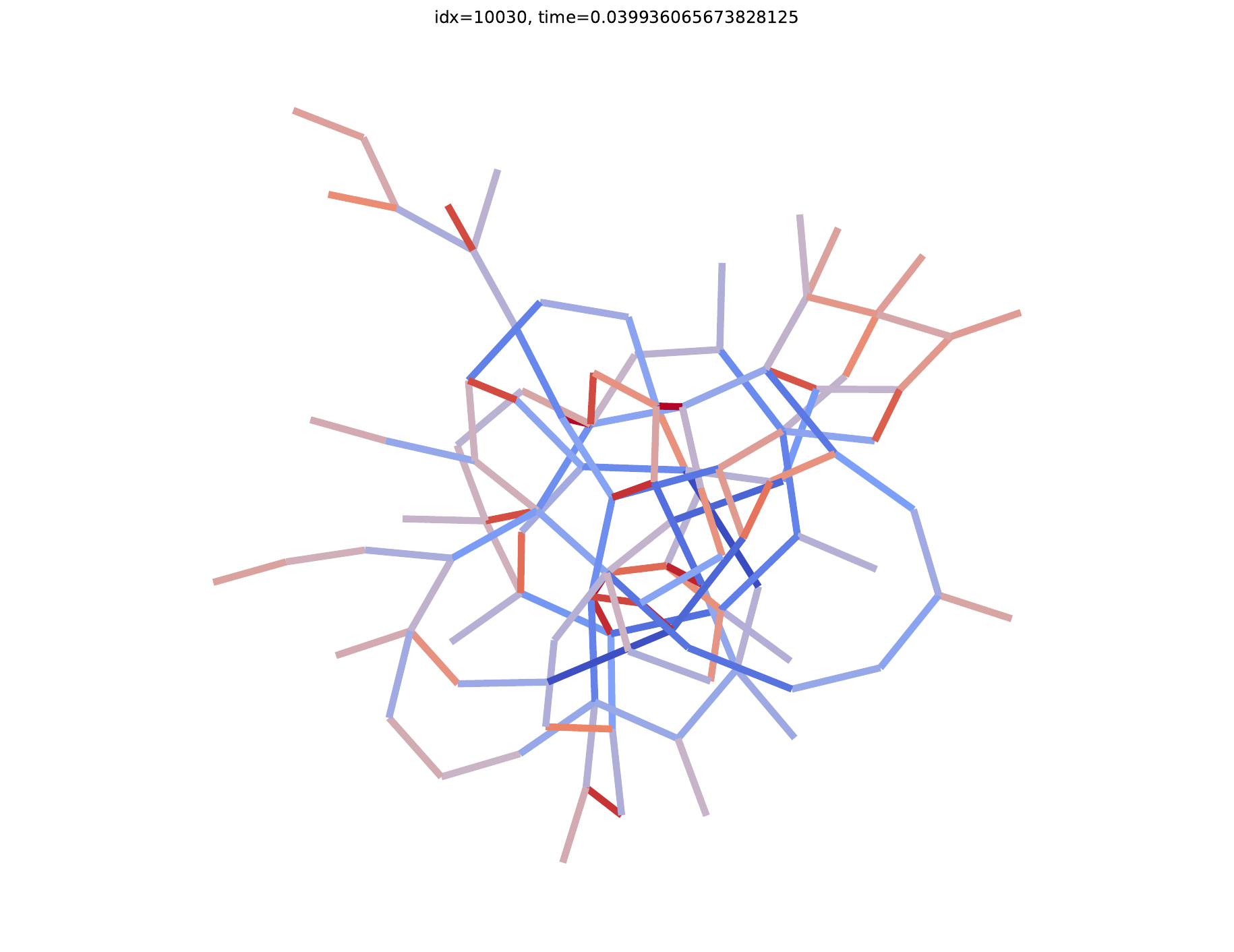} &
\imgcell{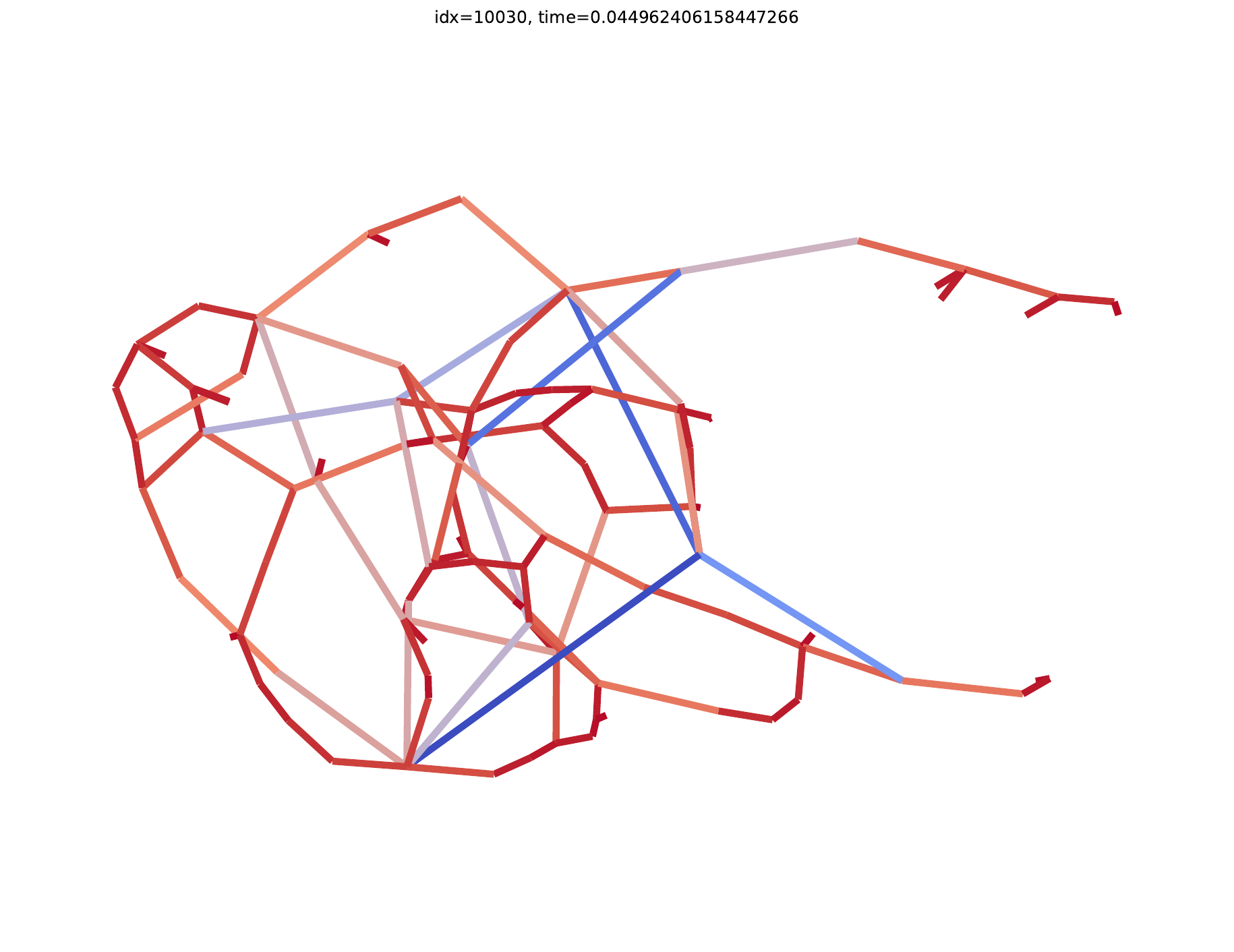} &
\imgcell{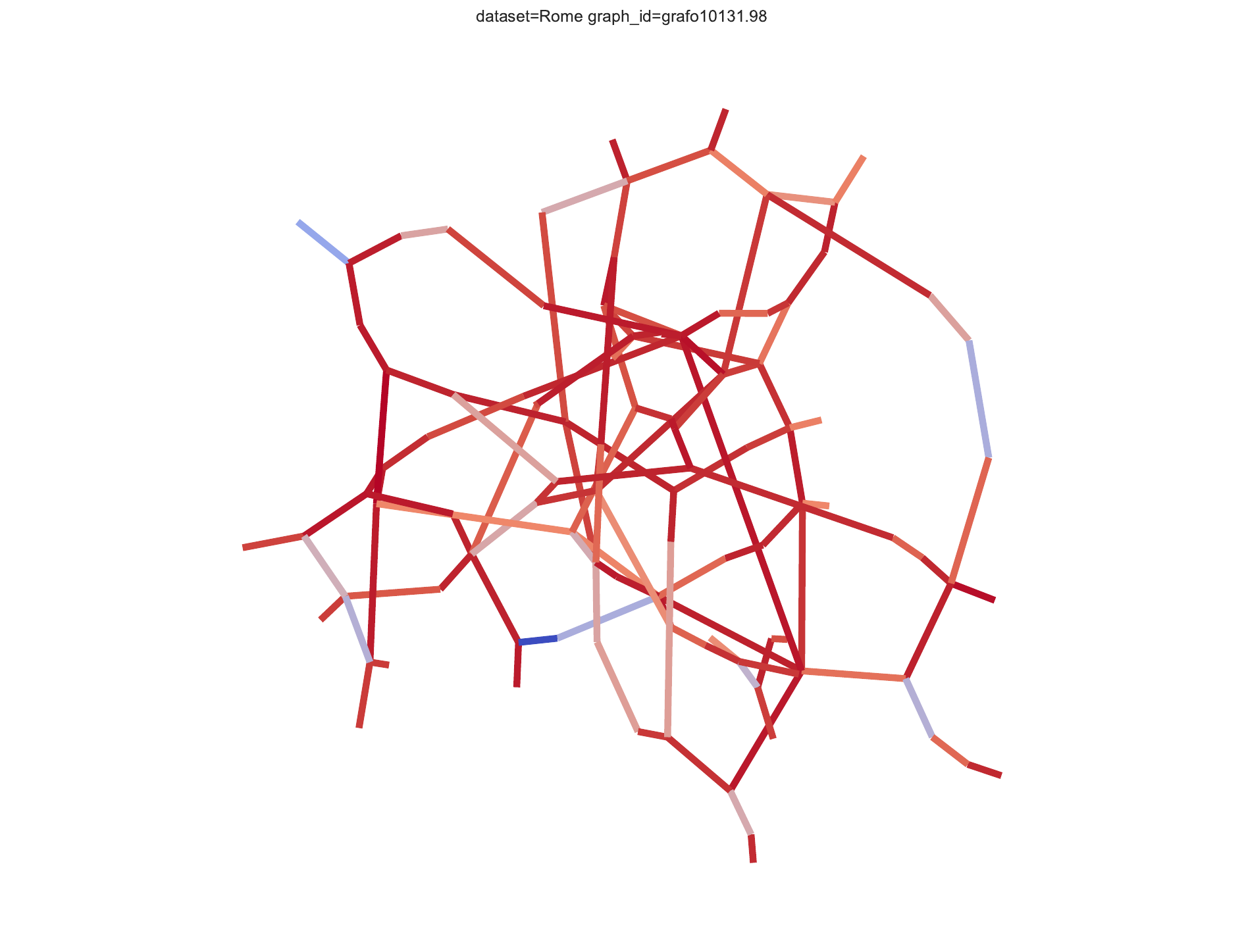} &
\imgcell{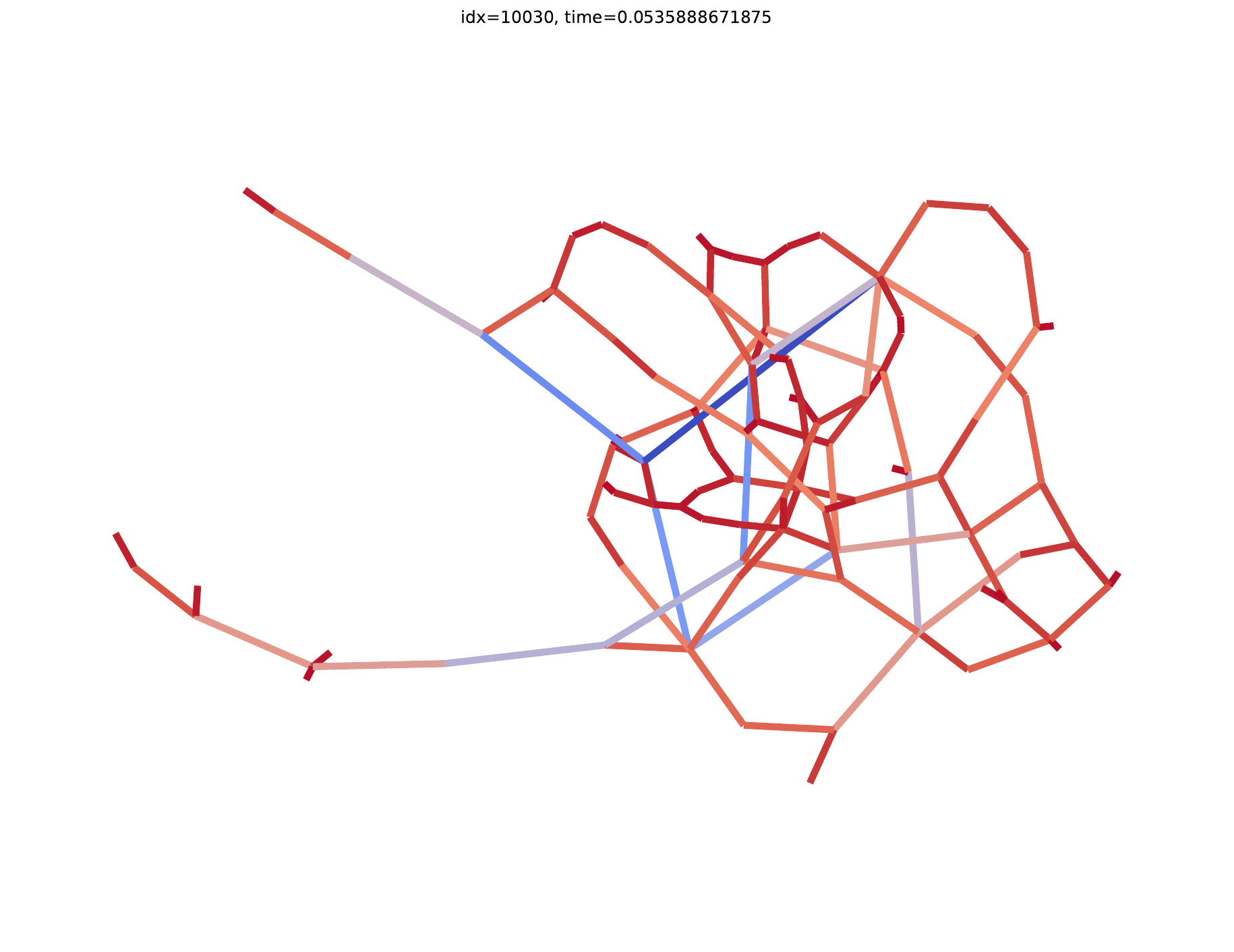} &
\imgcell{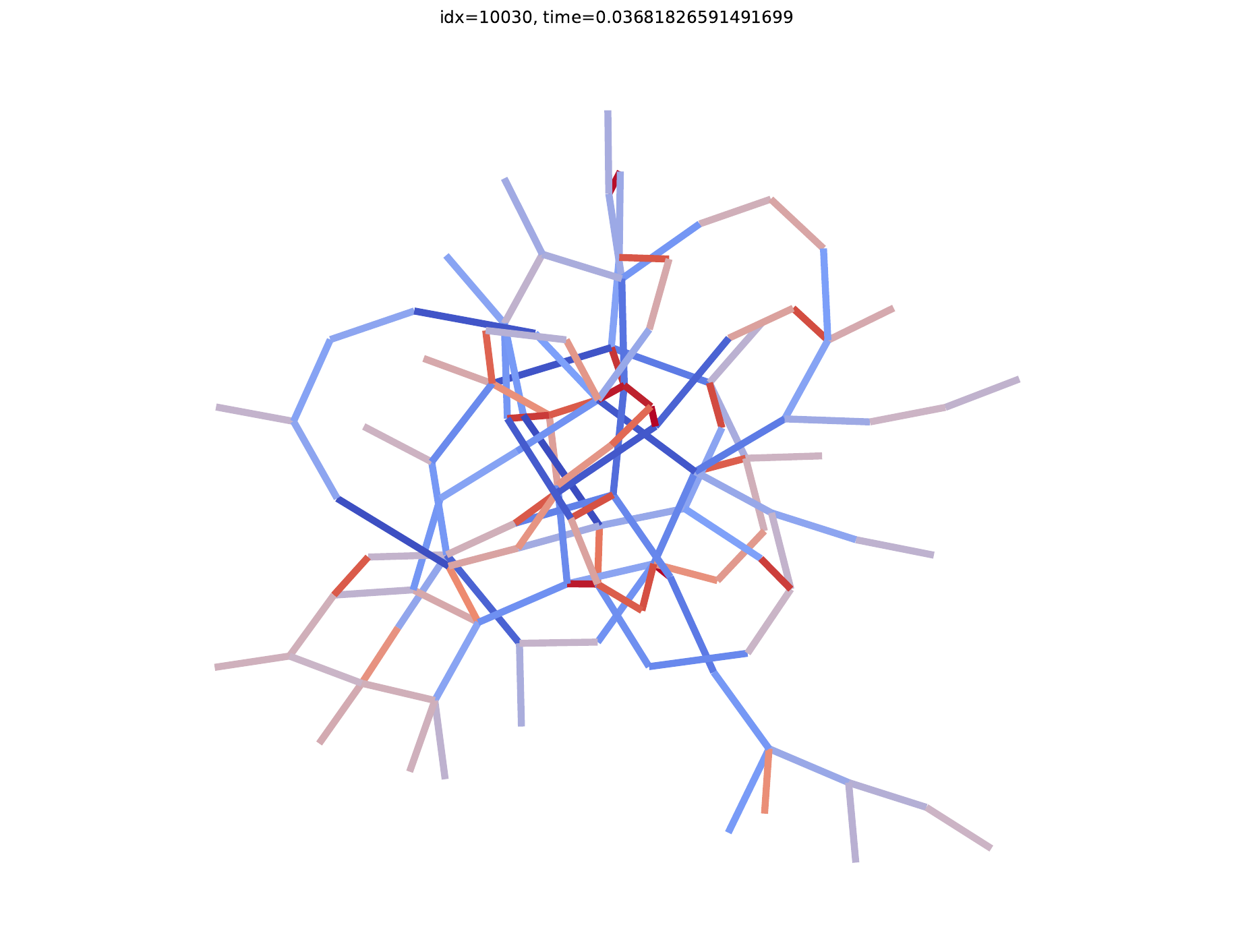} &
\imgcell{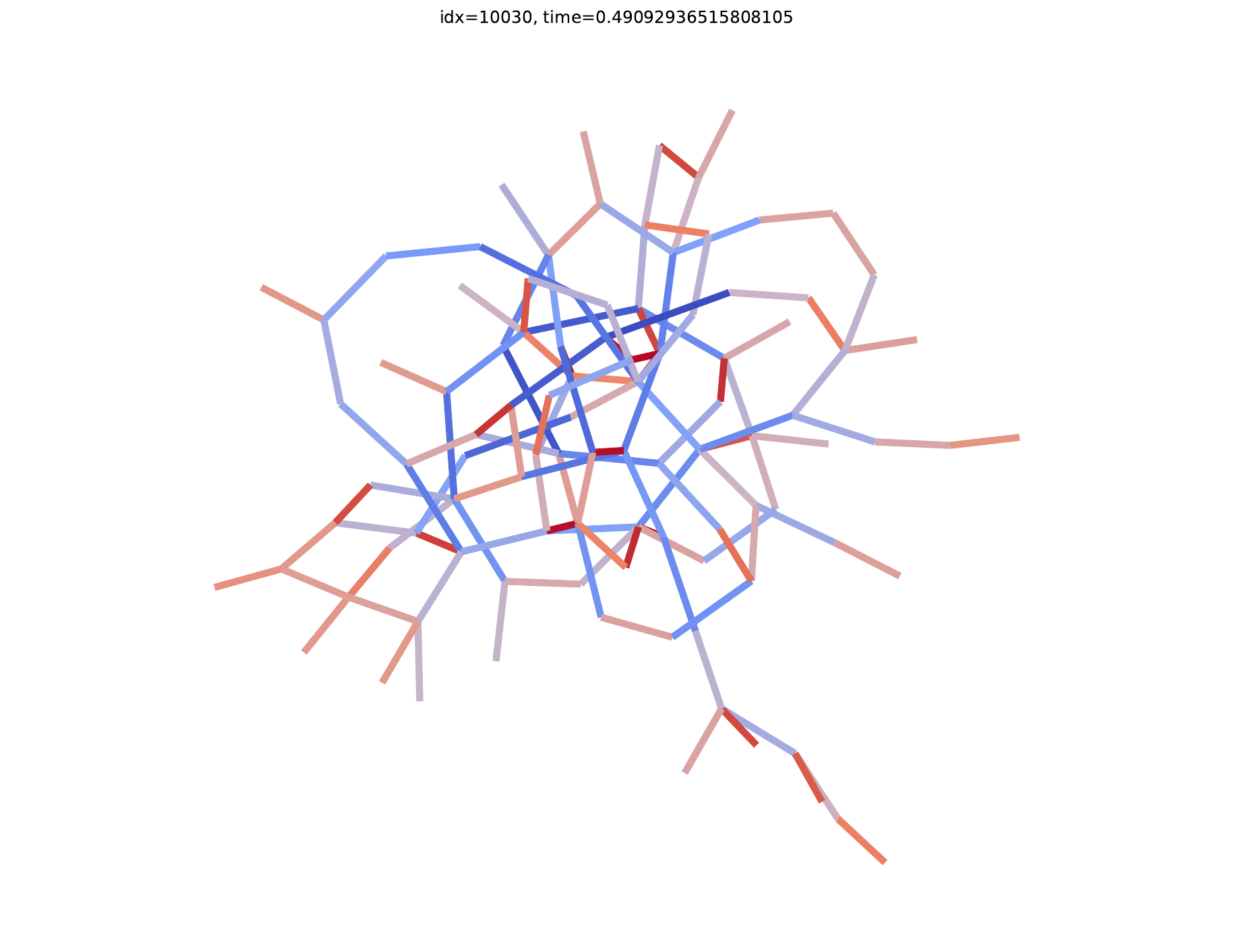} &
\imgcell{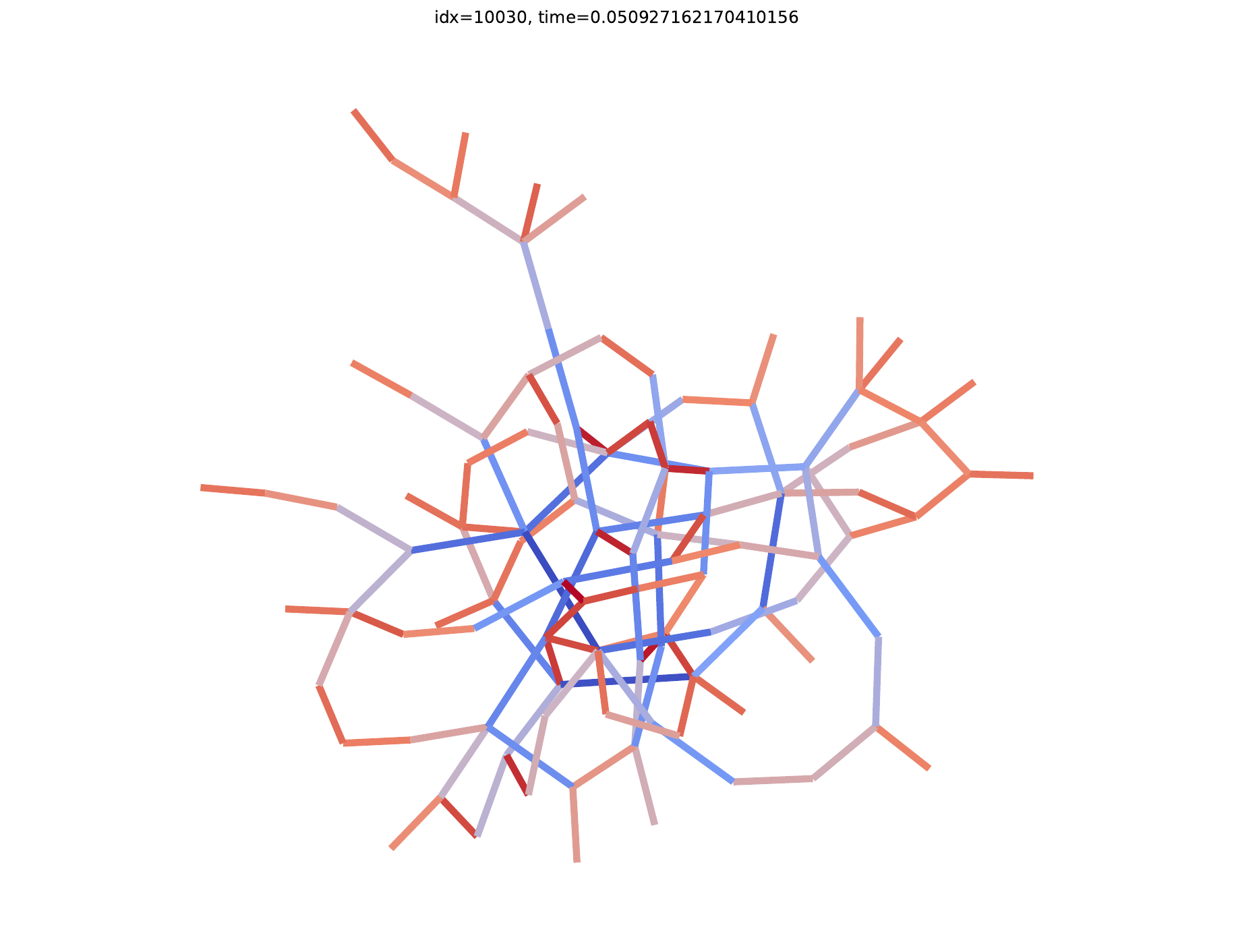} \\

&
t = 0.00s &
t = 1.66s &
t = 0.69s &
t = 0.05s &
t = 102.81s &
t = 0.04s &
t = 0.04s &
t = 0.04s &
t = 0.05s &
t = 0.04s &
t = 0.06s &
t = 0.05s \\

\makecell{\bfseries grafo7261.42\\N = 49\\M = 57} &
\imgcell{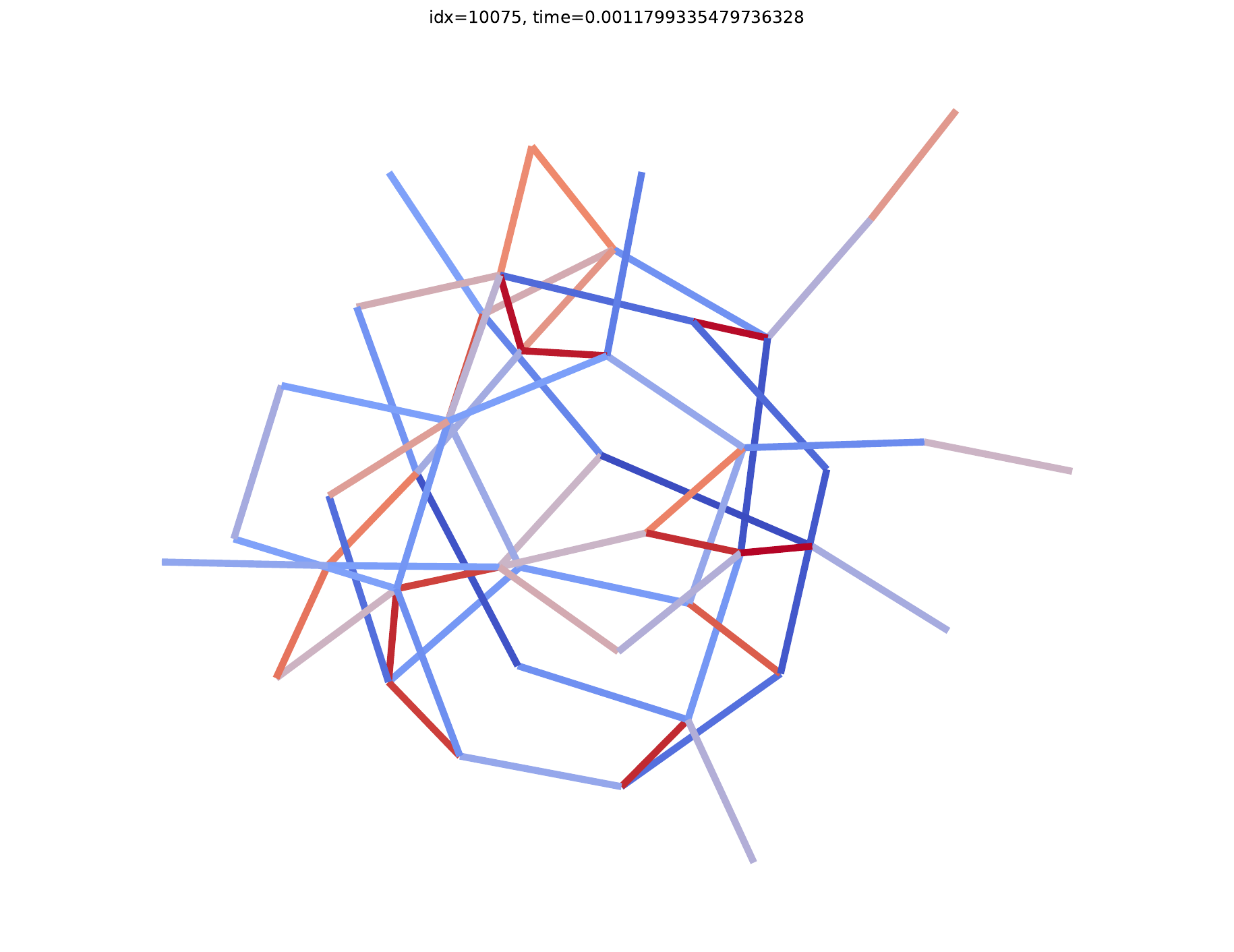} &
\imgcell{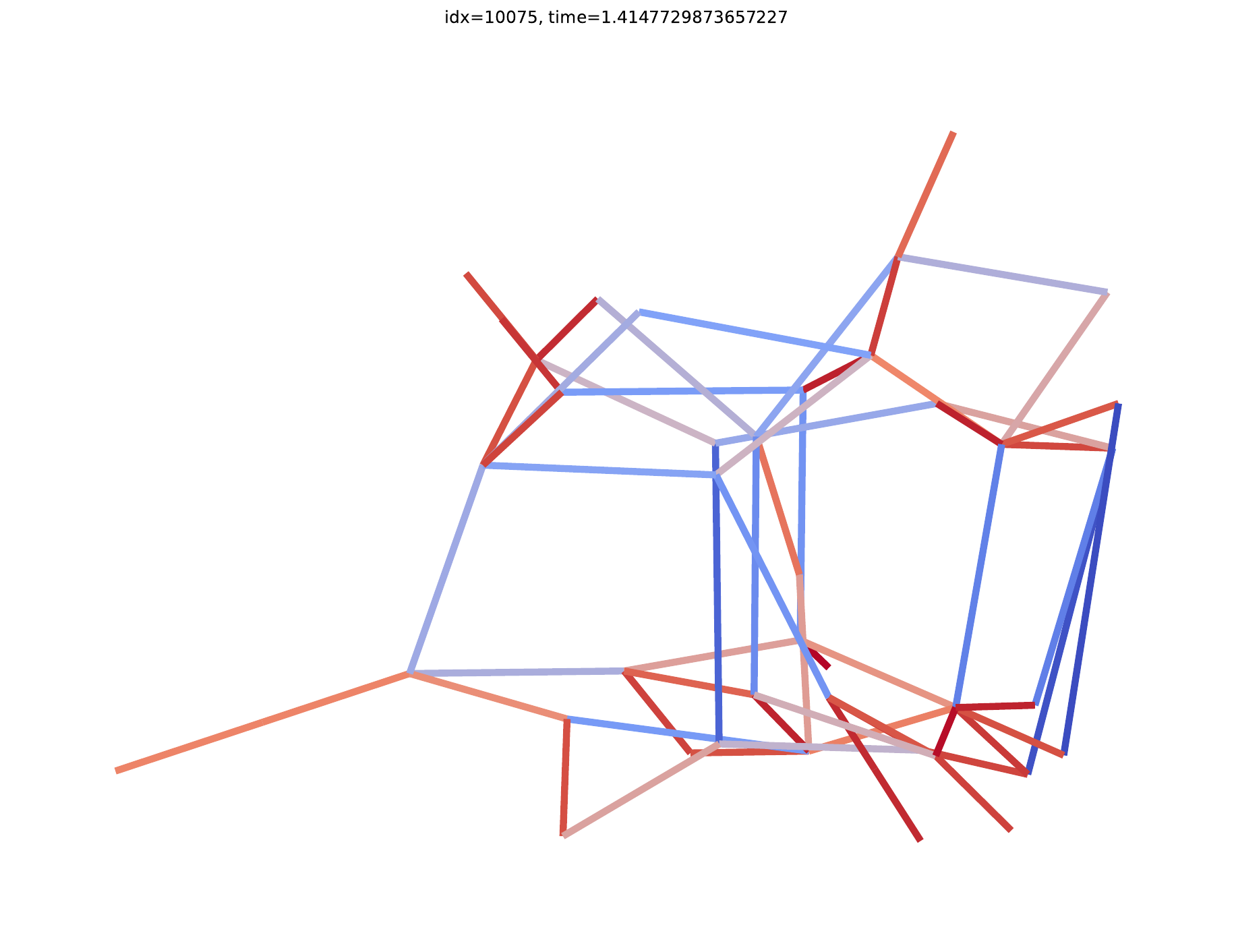} &
\imgcell{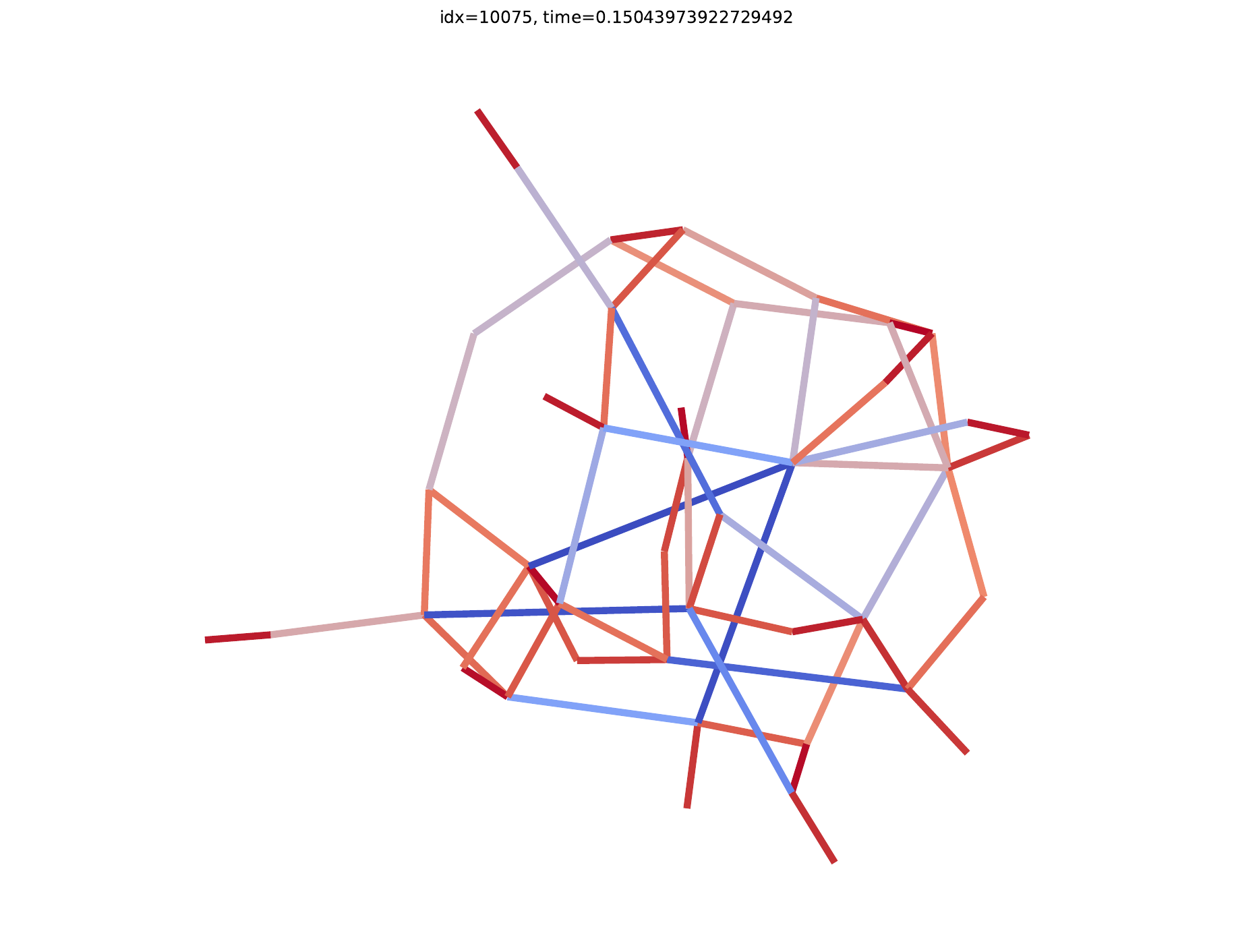} &
\imgcell{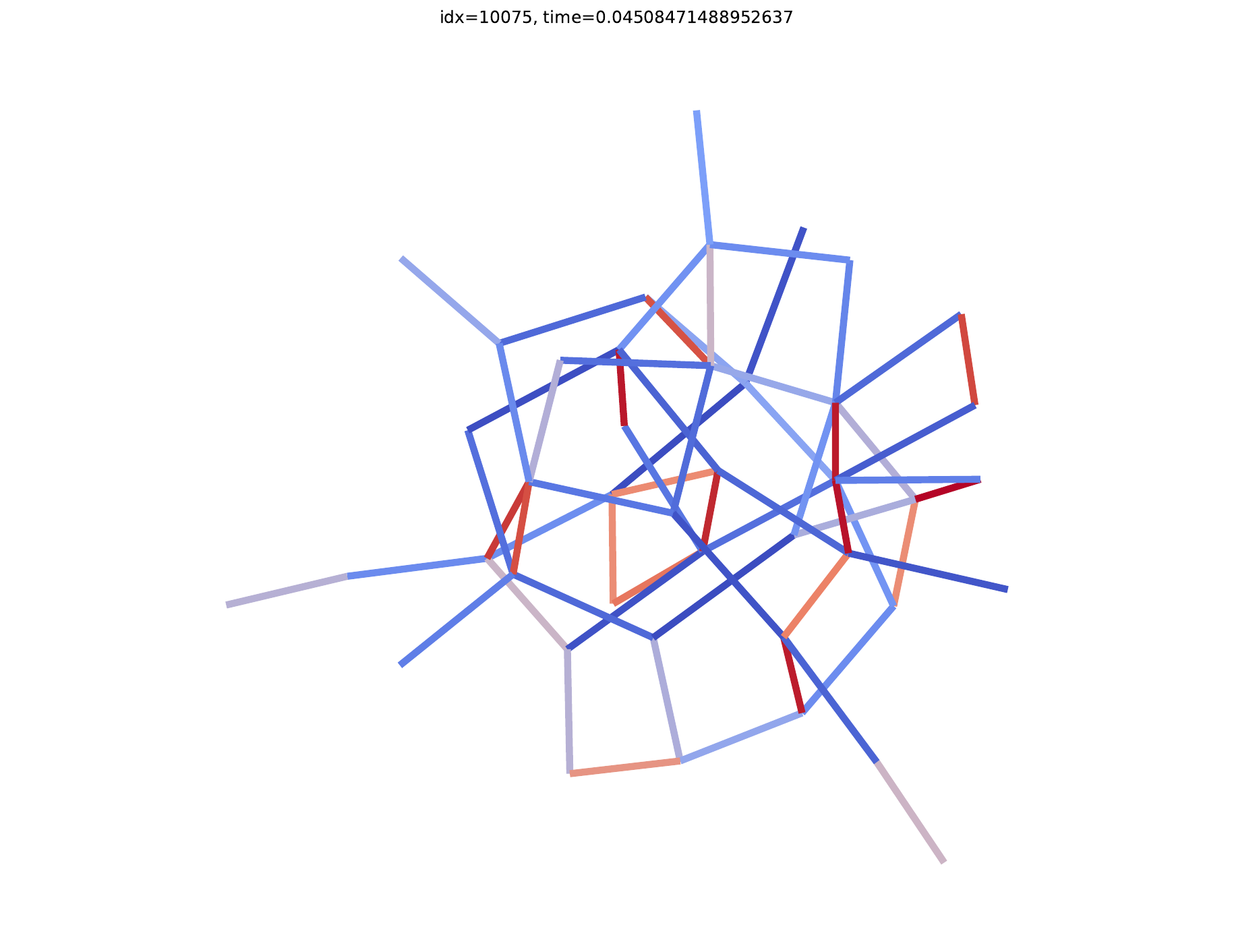} &
\imgcell{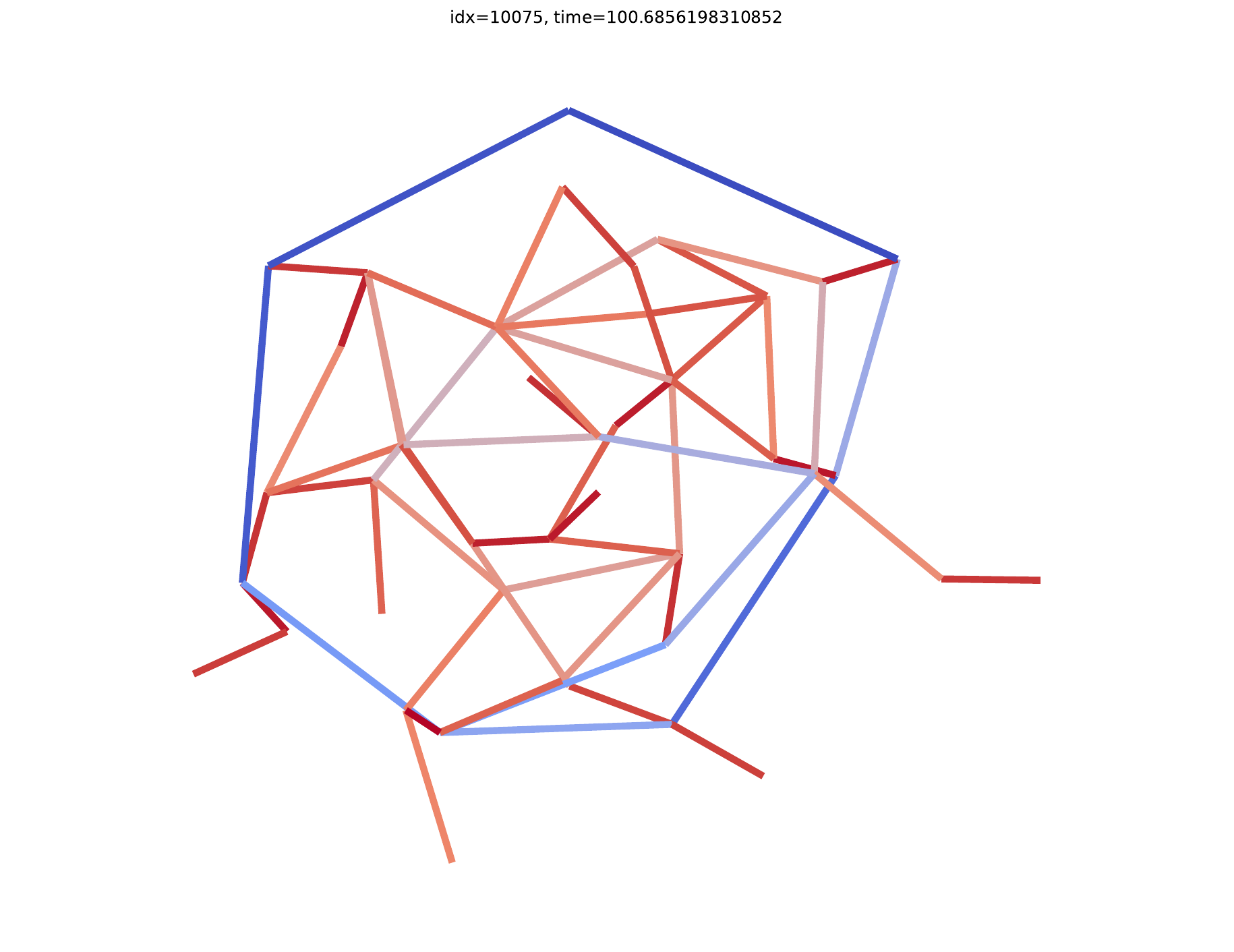} &
\imgcell{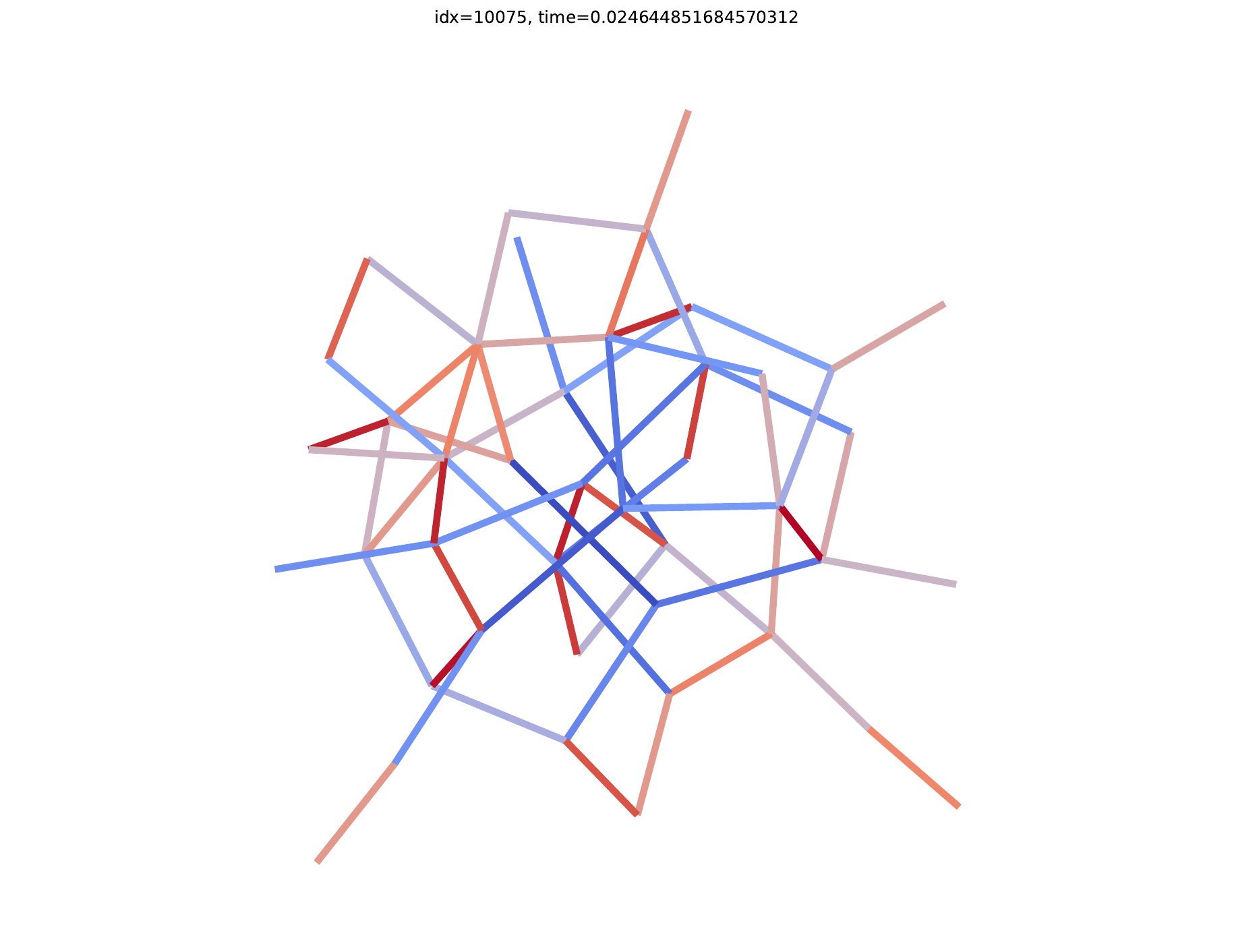} &
\imgcell{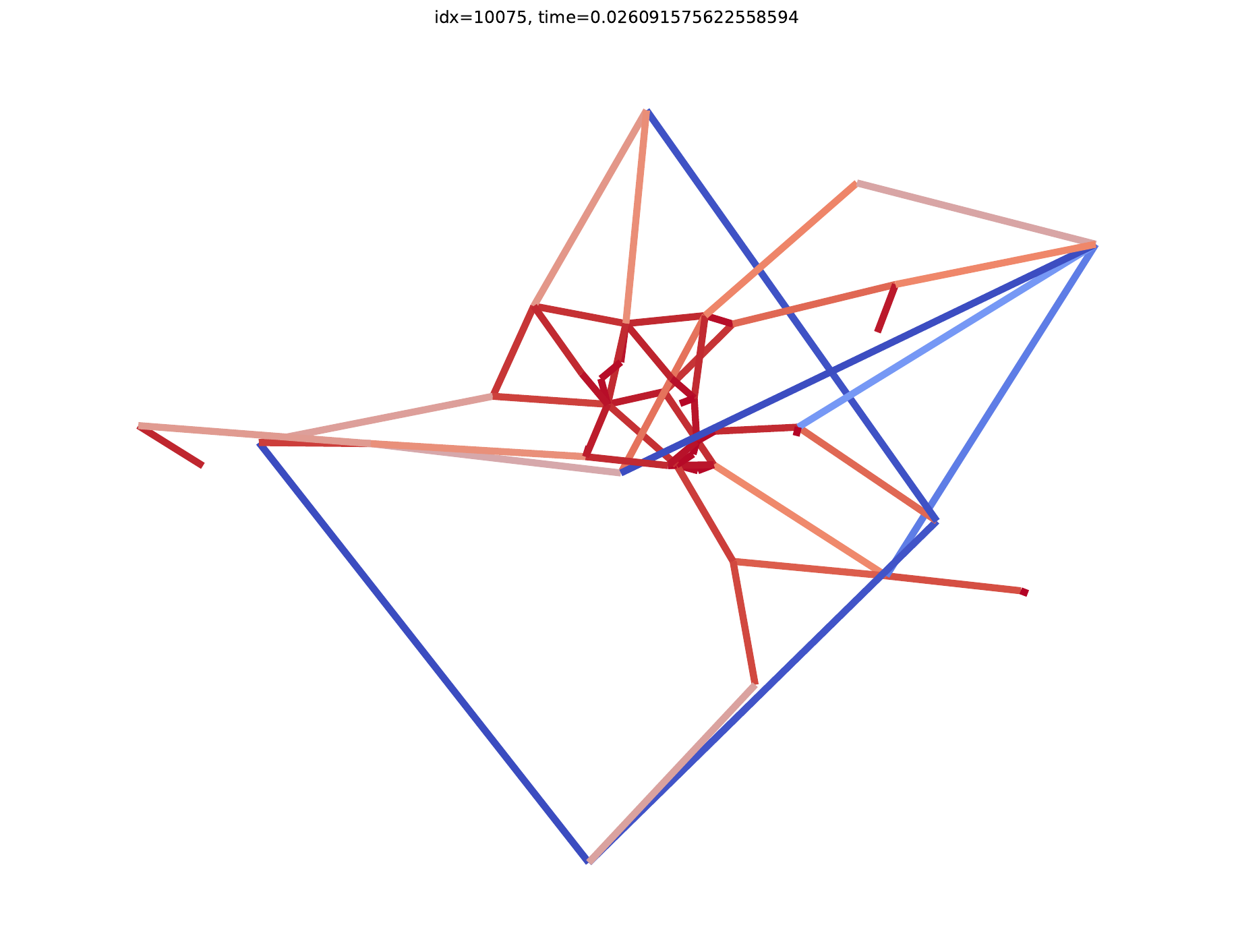} &
\imgcell{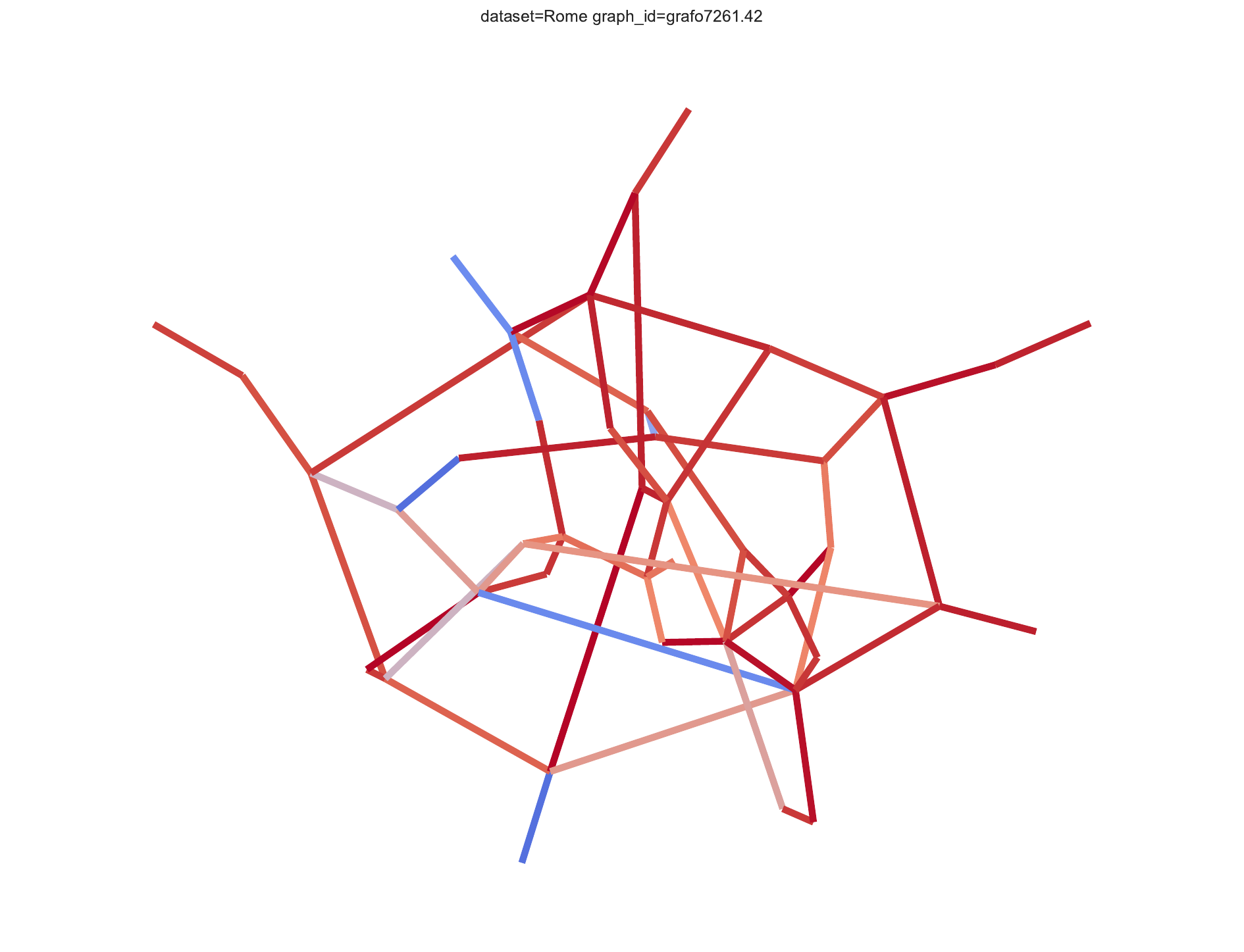} &
\imgcell{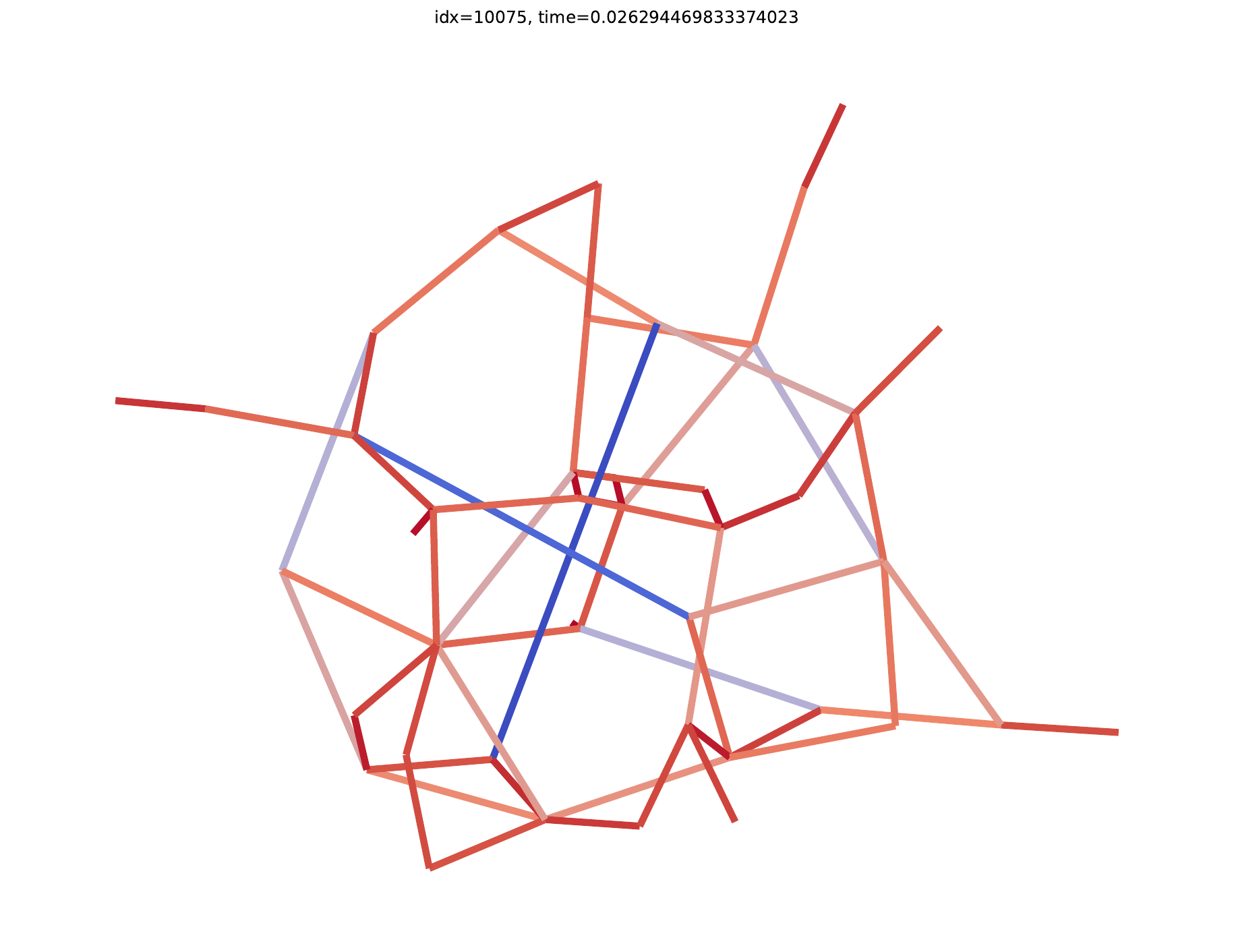} &
\imgcell{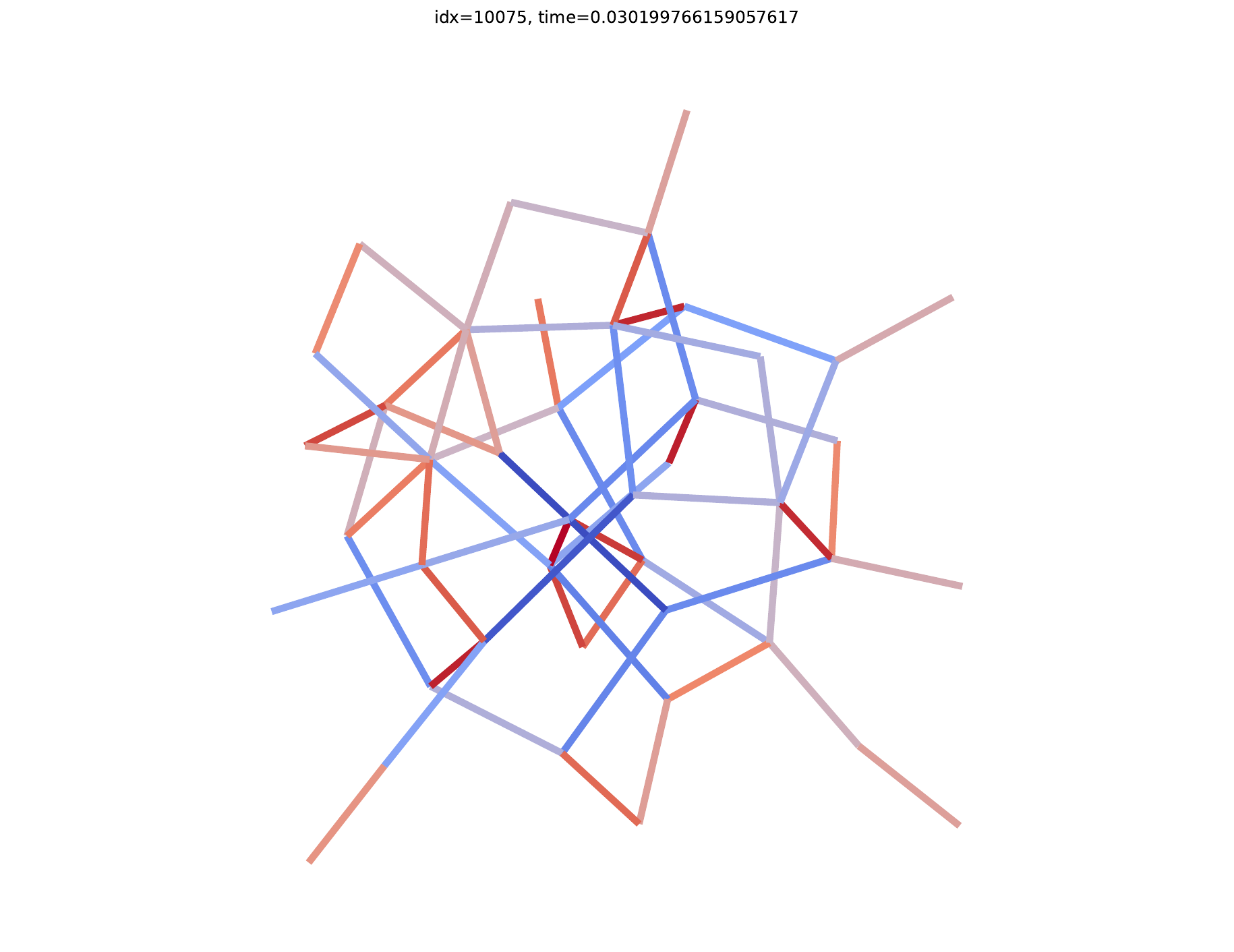} &
\imgcell{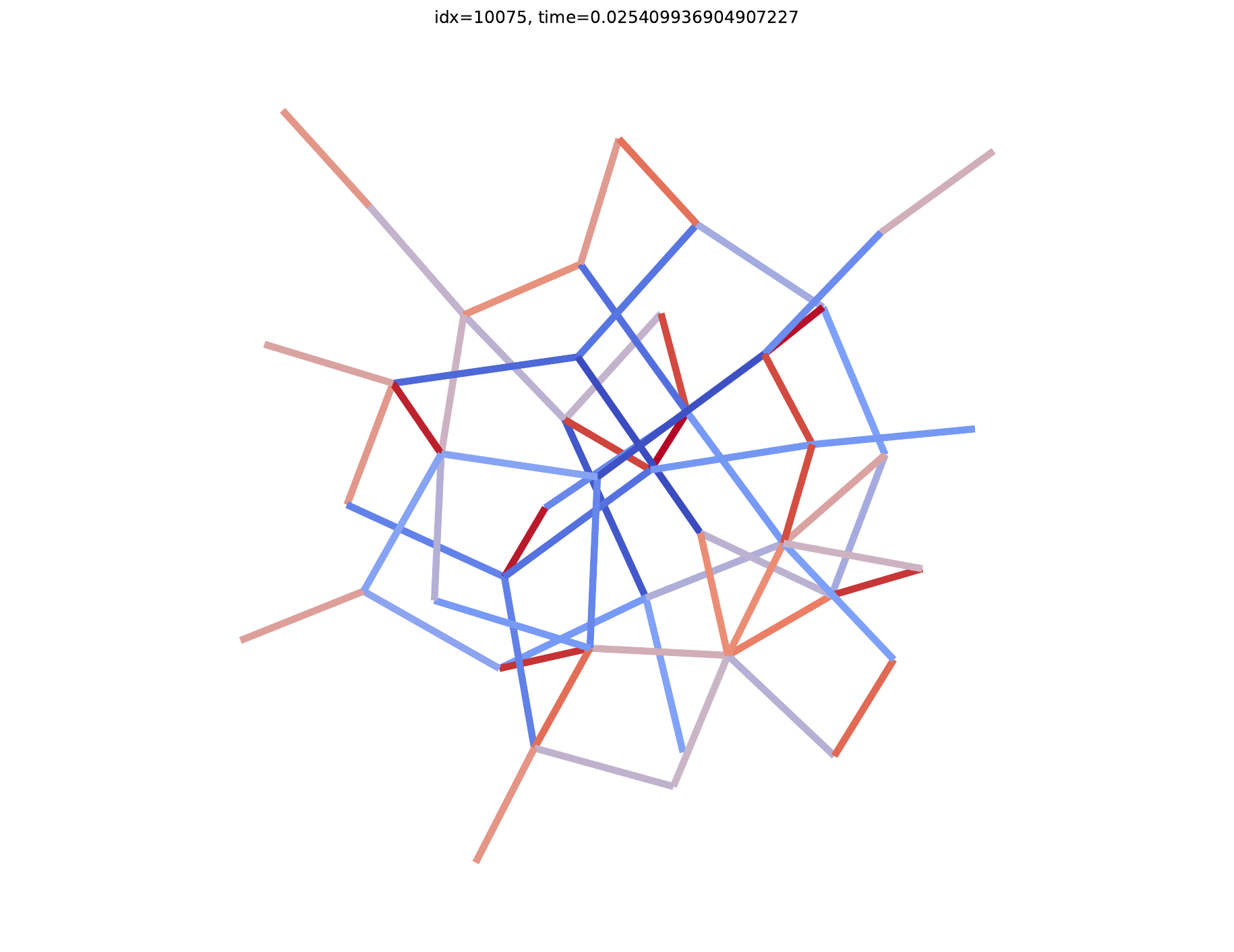} &
\imgcell{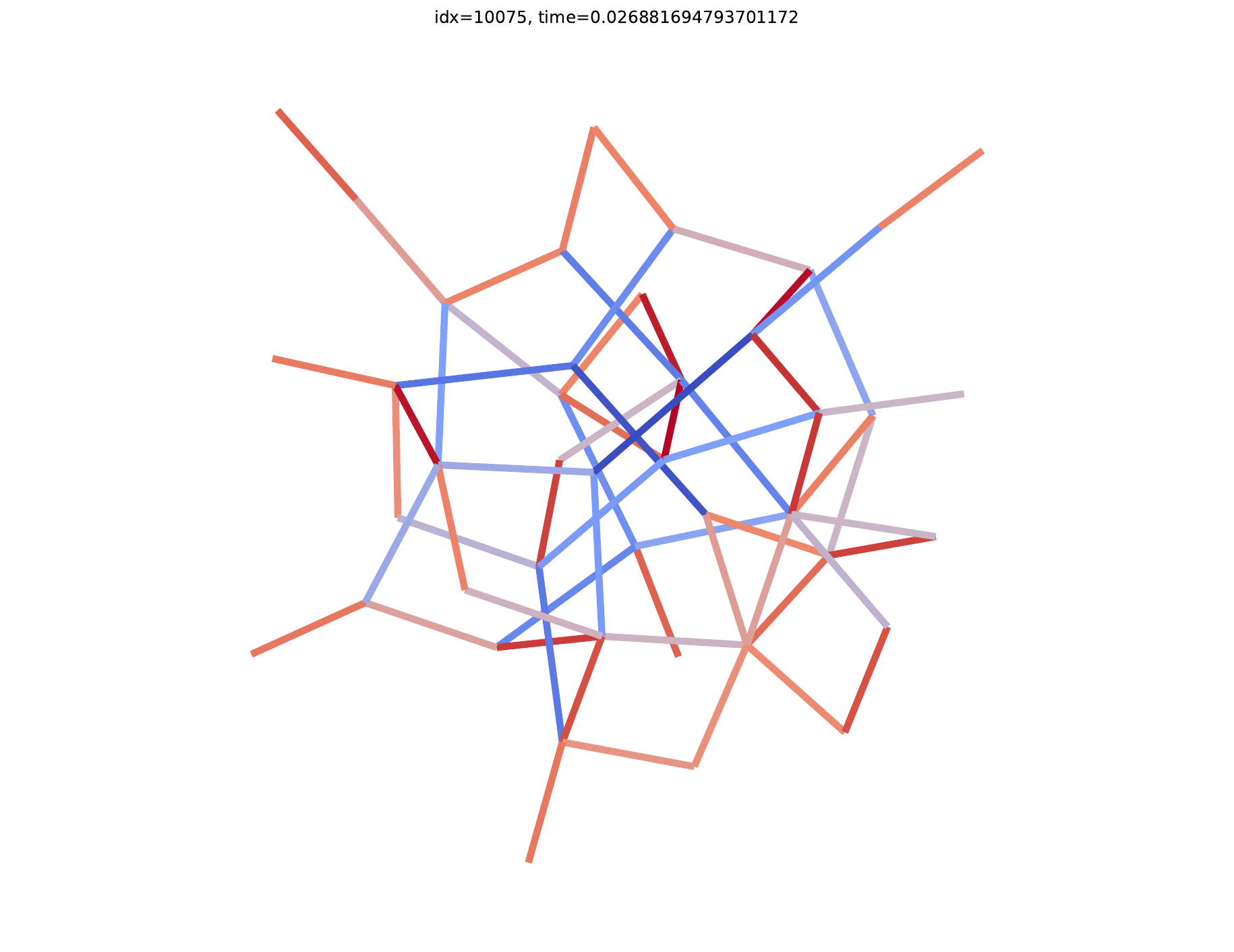} \\

&
t = 0.00s &
t = 1.41s &
t = 0.15s &
t = 0.05s &
t = 100.69s &
t = 0.02s &
t = 0.03s &
t = 0.04s &
t = 0.03s &
t = 0.03s &
t = 0.03s &
t = 0.03s \\

\makecell{\bfseries grafo833.13\\N = 69\\M = 86} &
\imgcell{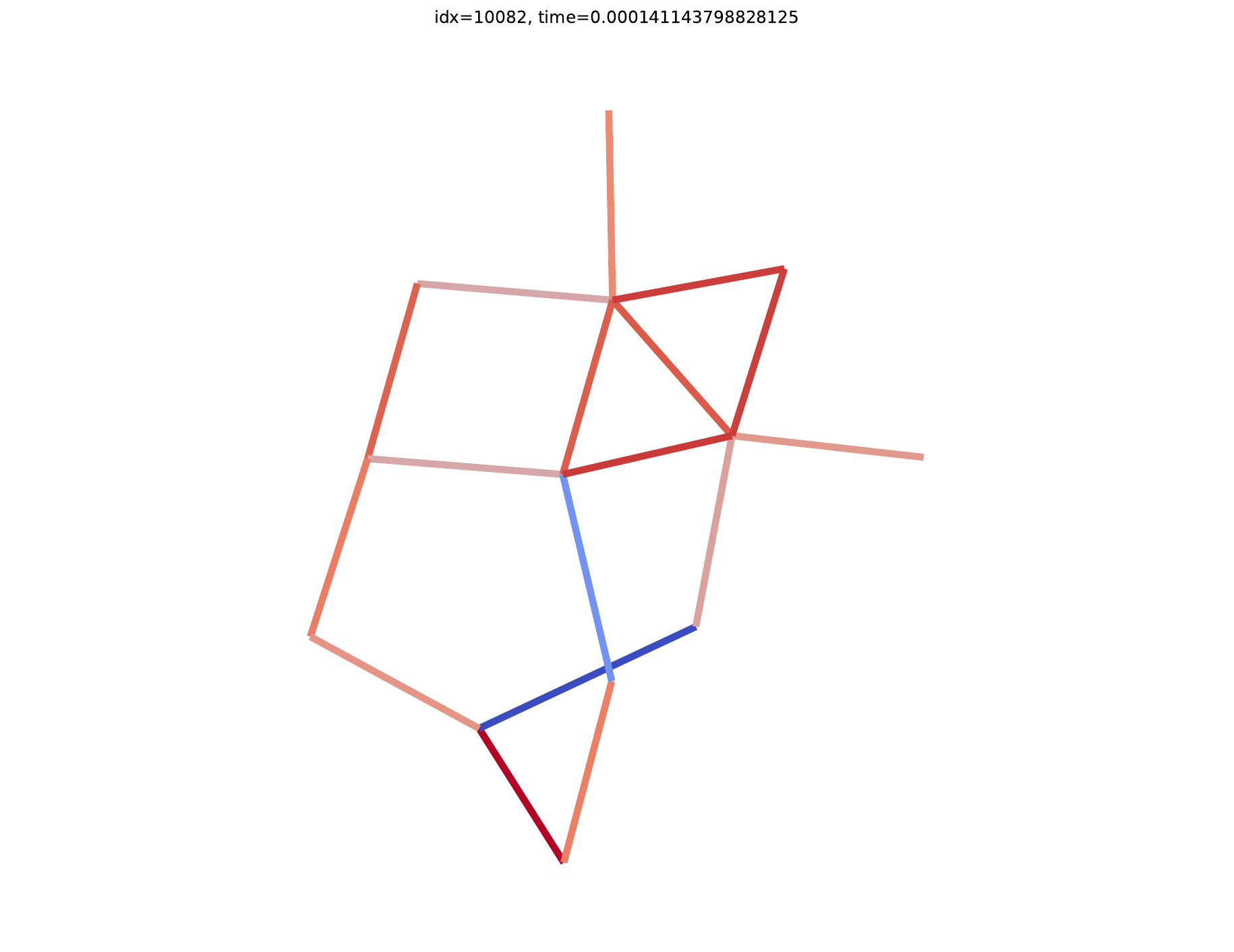} &
\imgcell{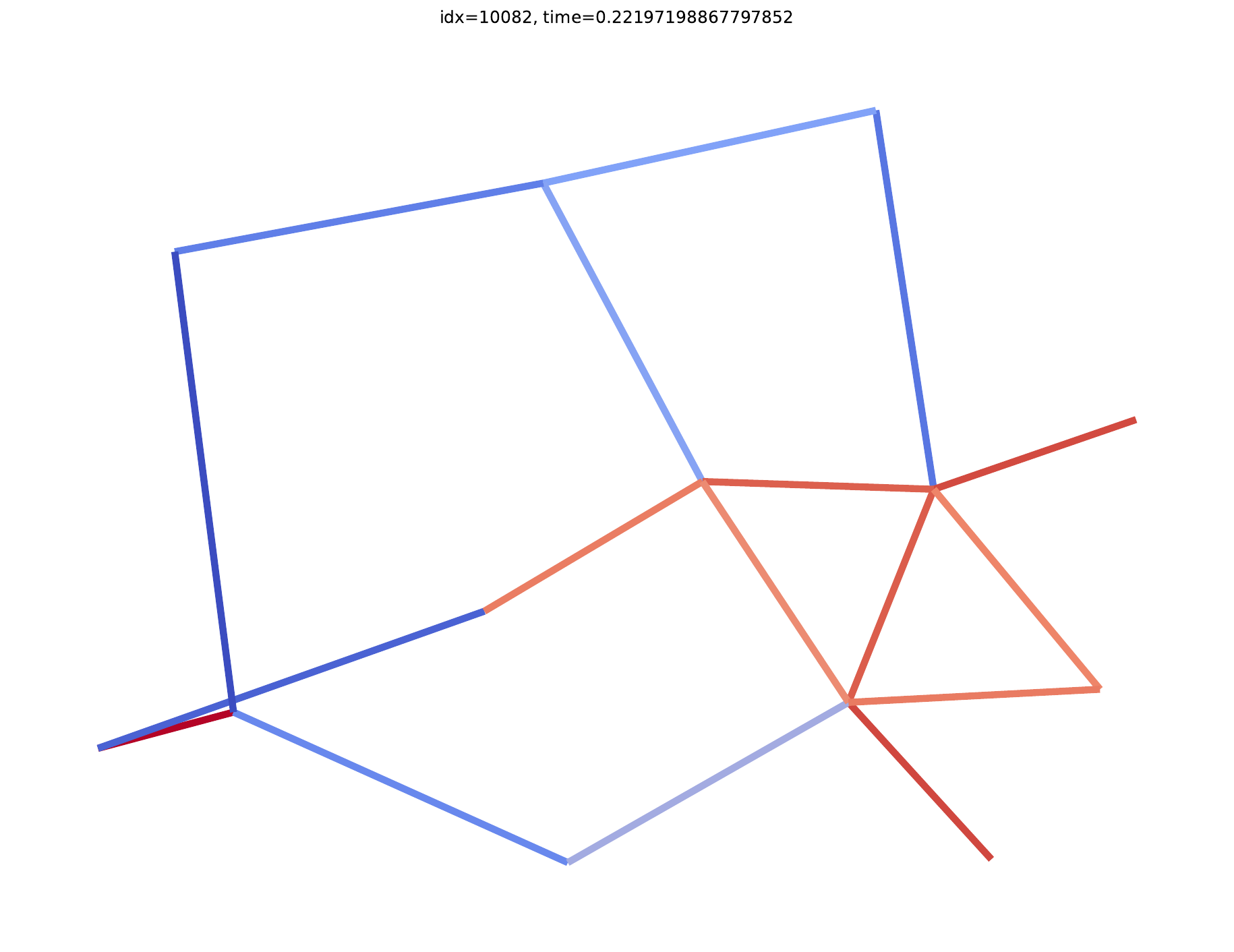} &
\imgcell{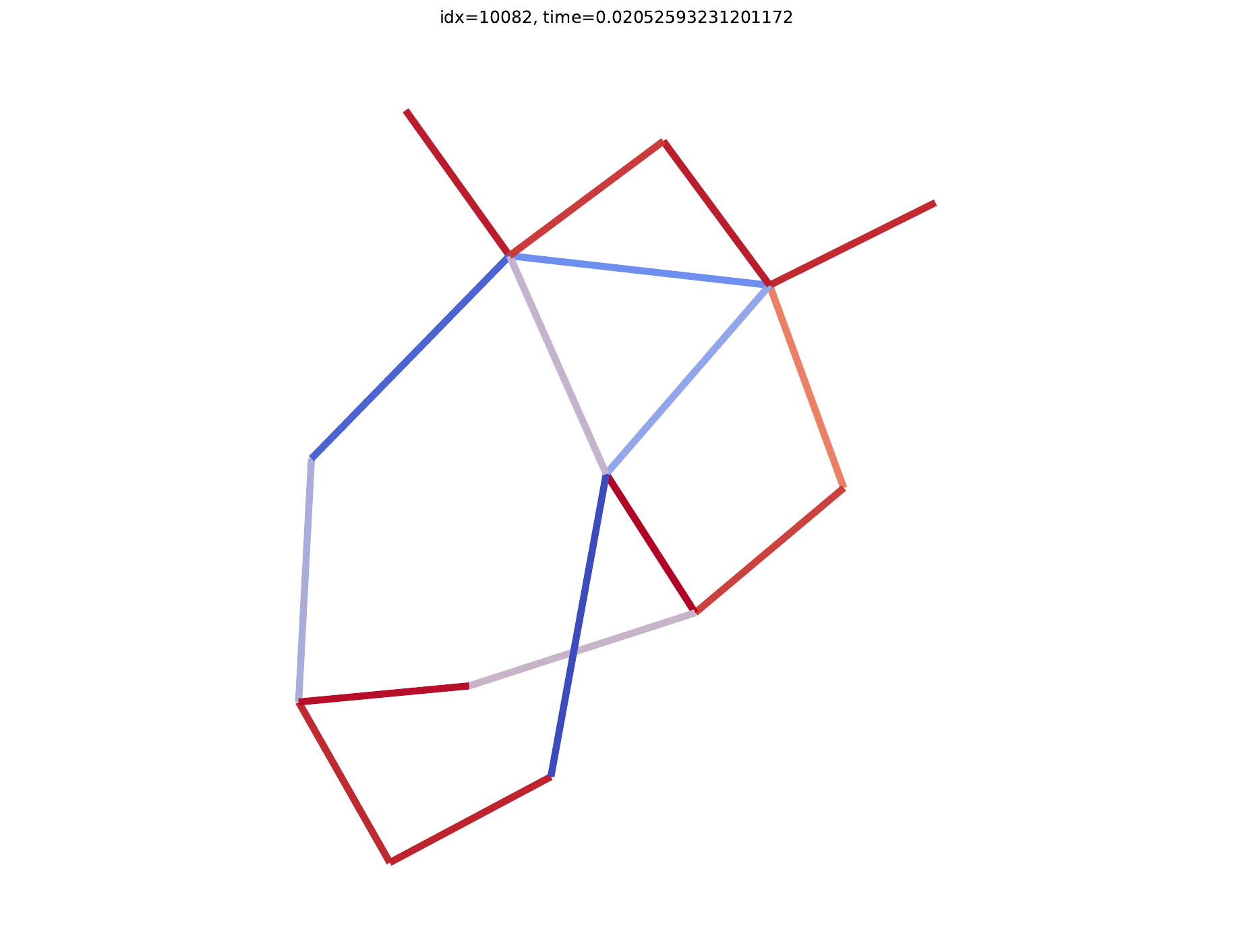} &
\imgcell{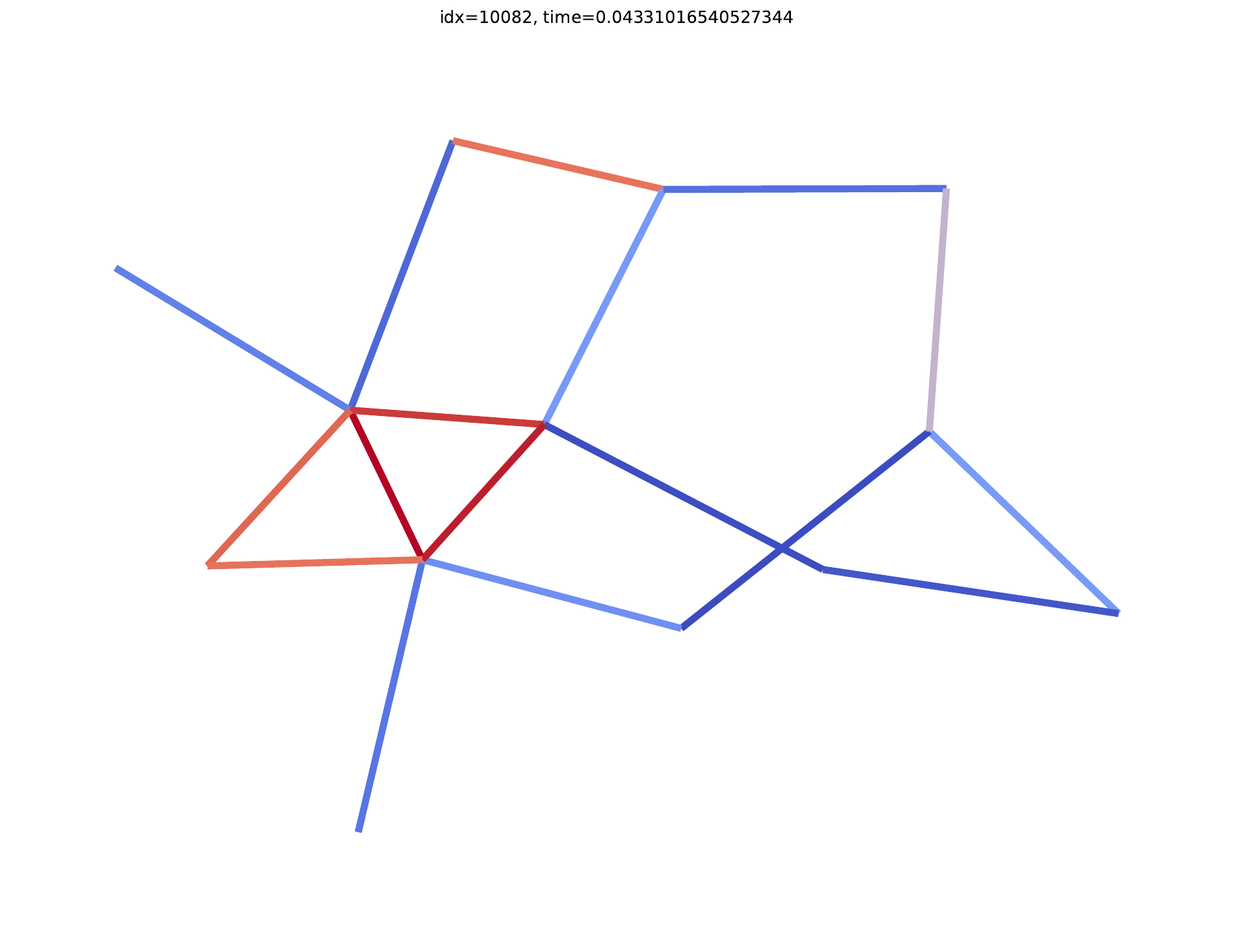} &
\imgcell{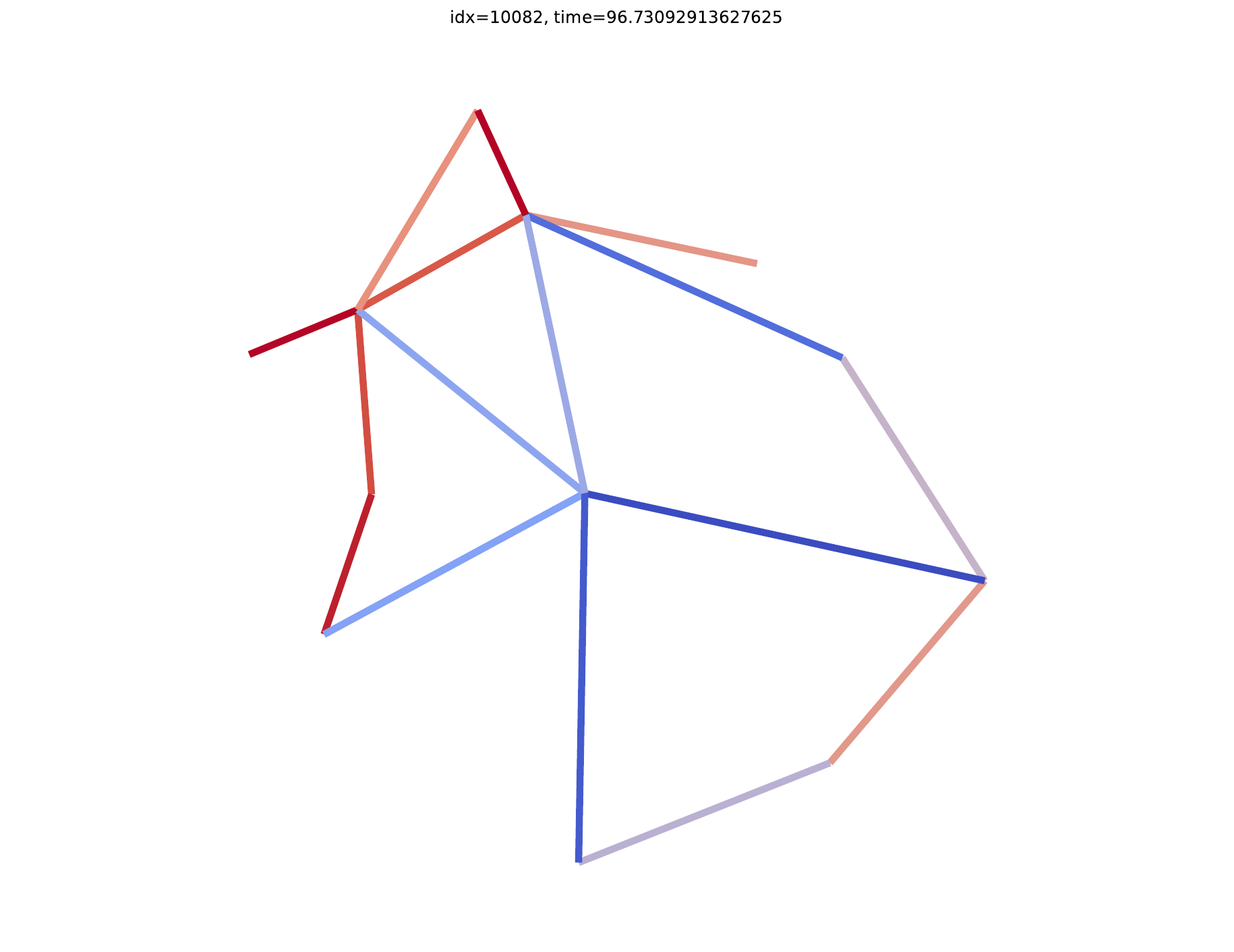} &
\imgcell{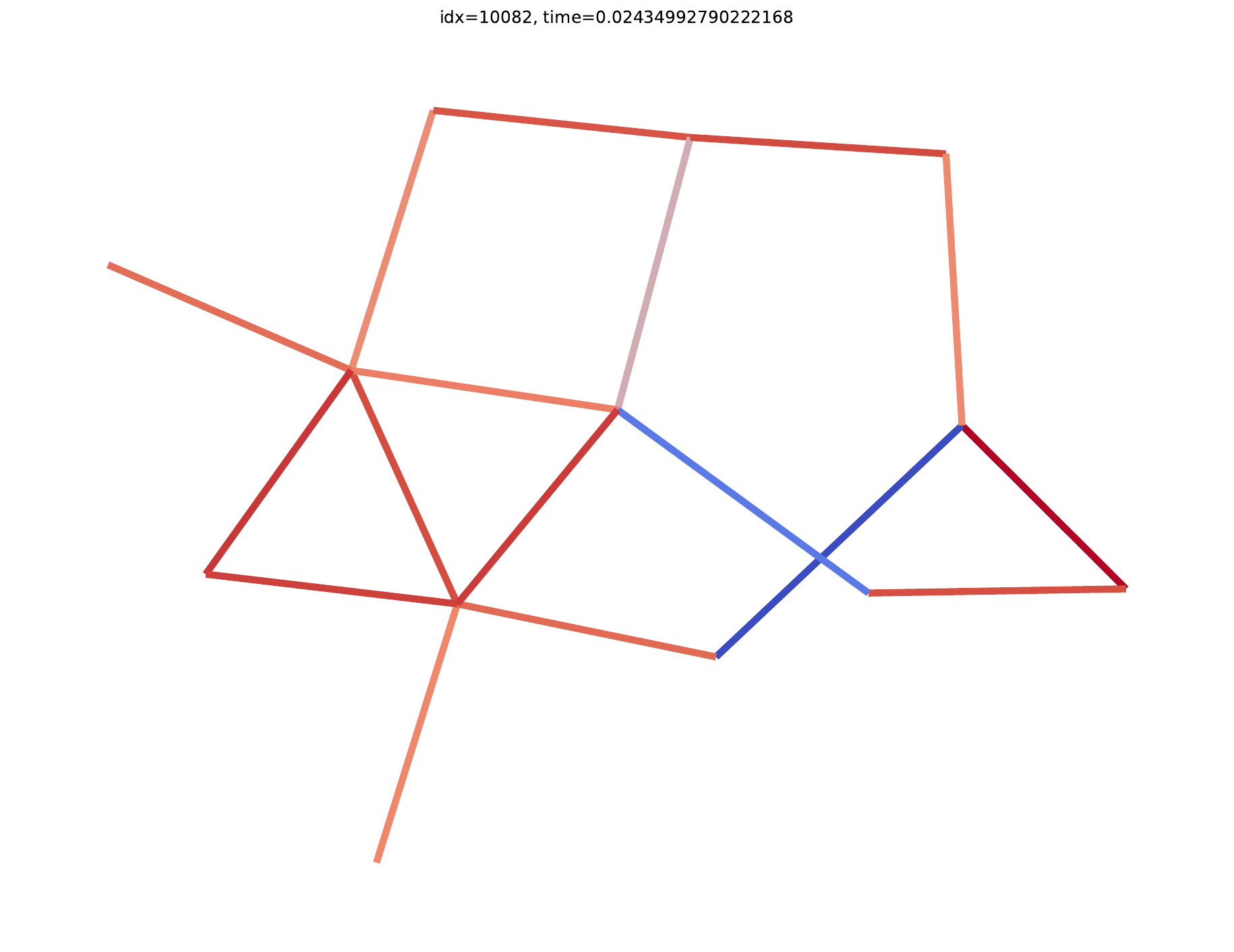} &
\imgcell{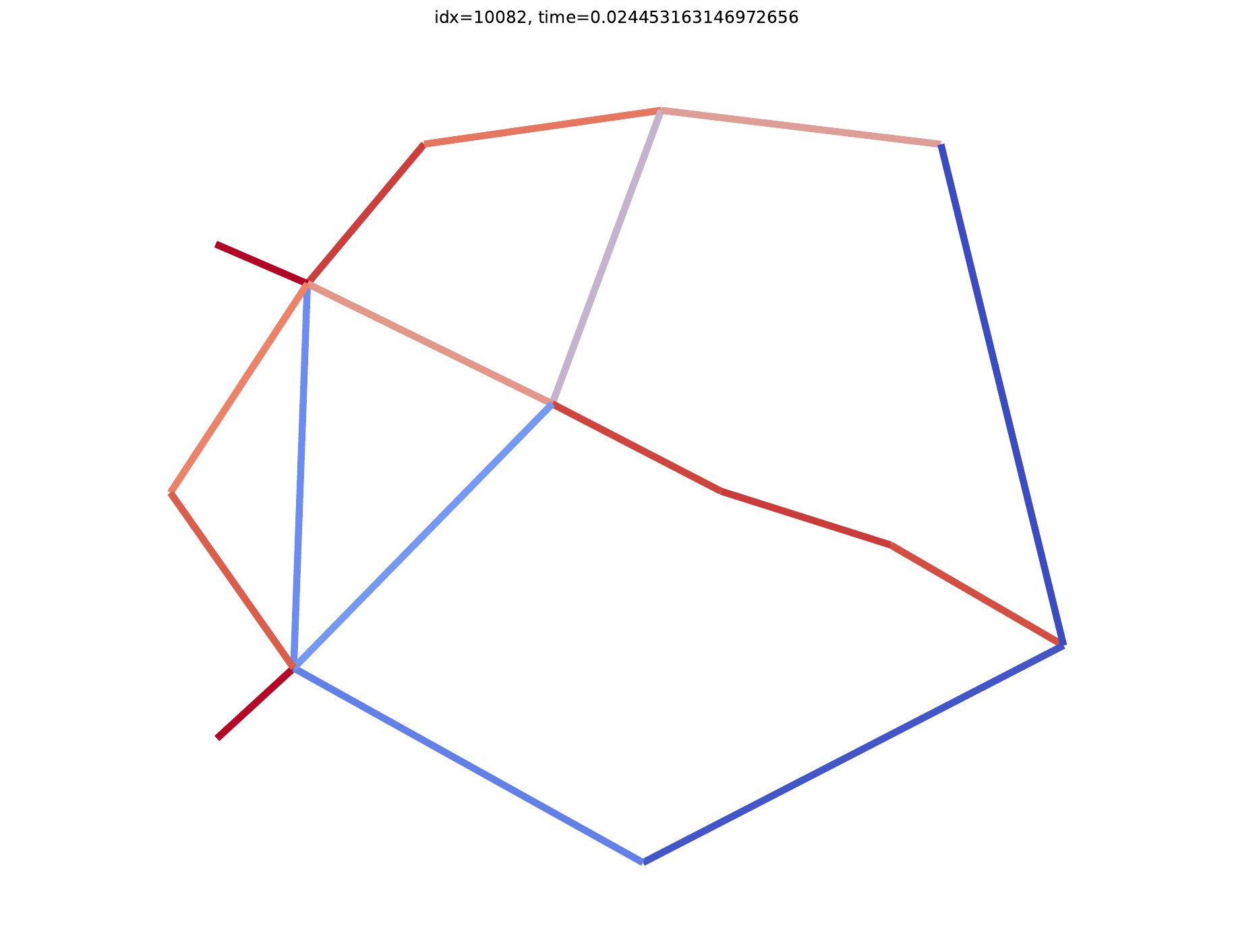} &
\imgcell{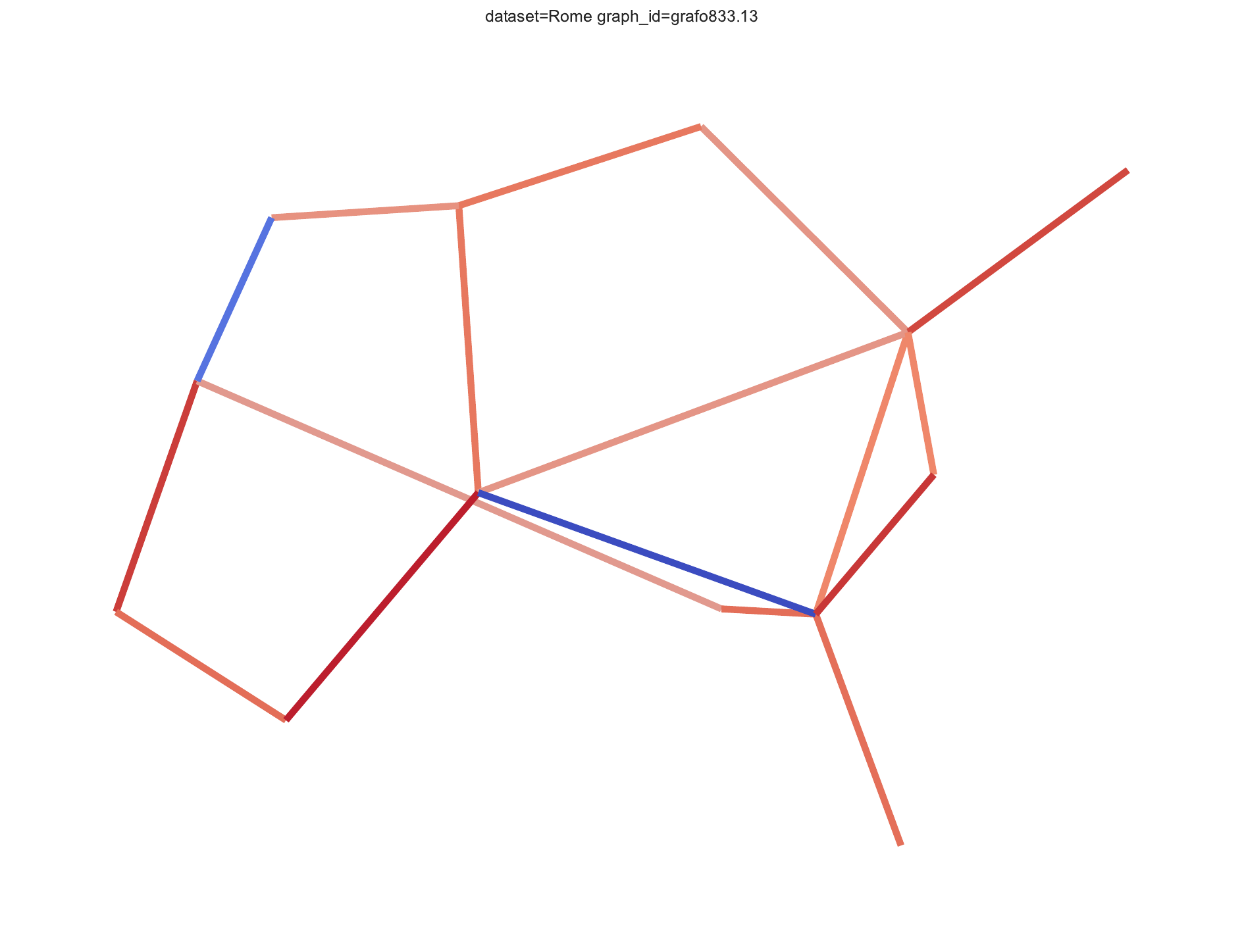} &
\imgcell{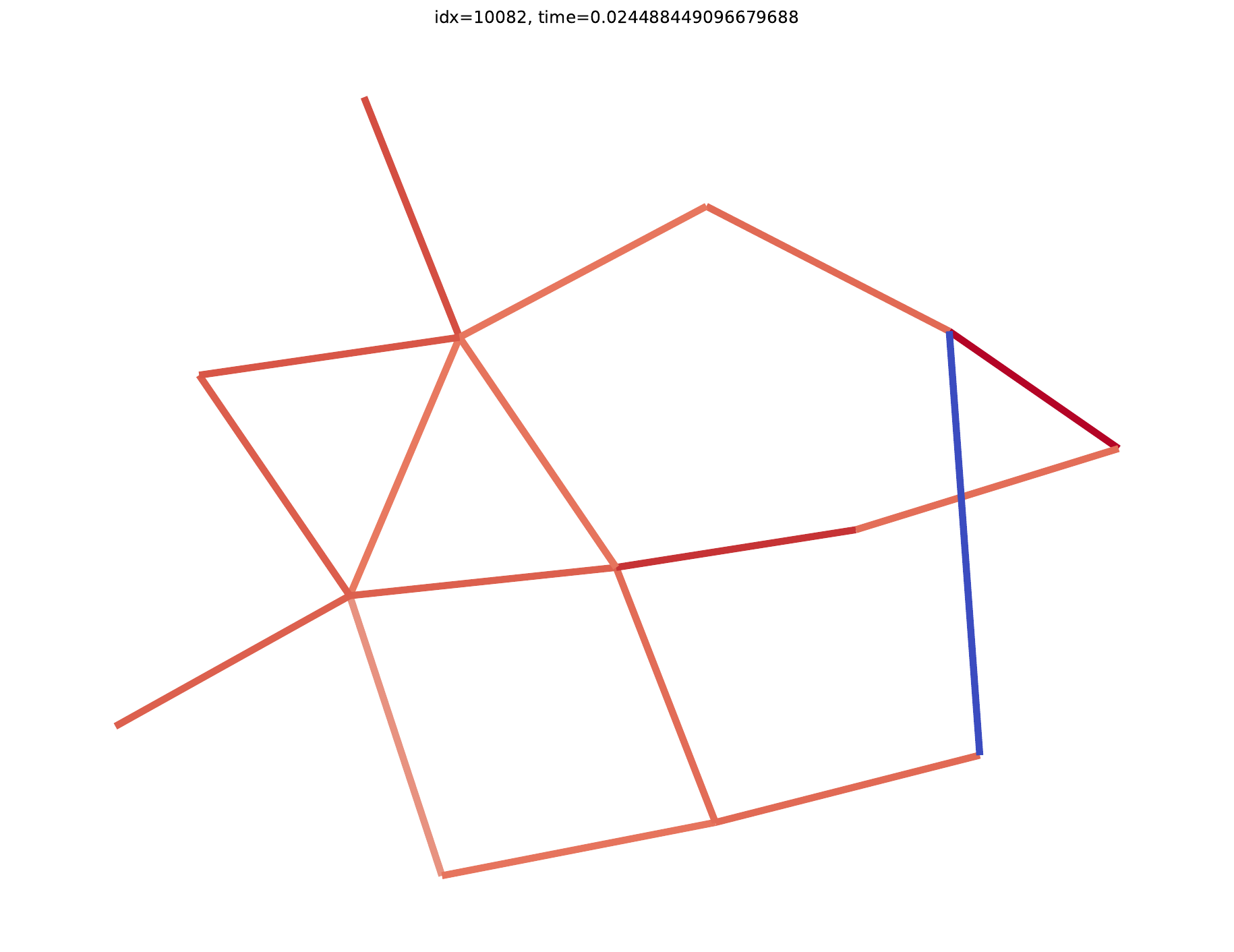} &
\imgcell{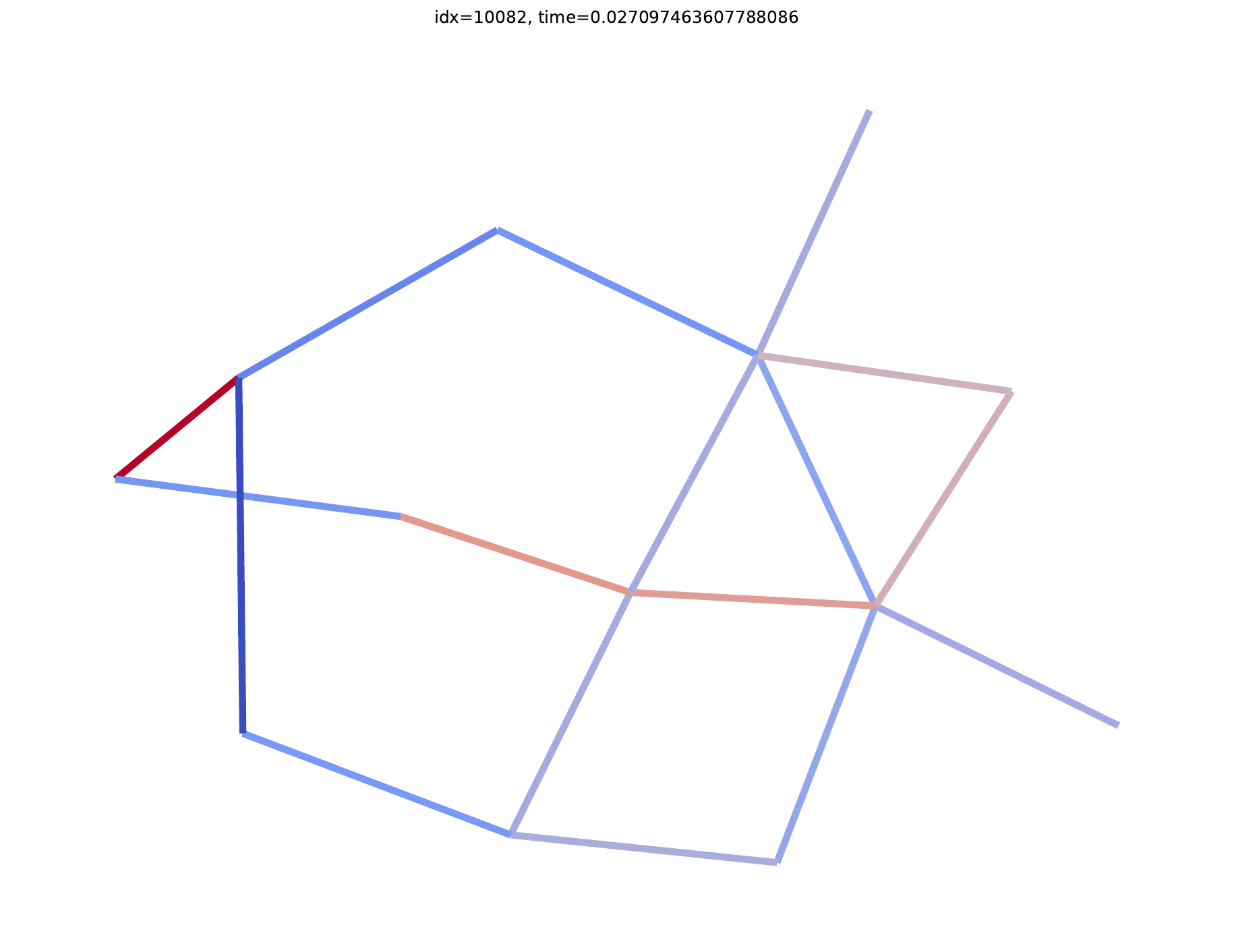} &
\imgcell{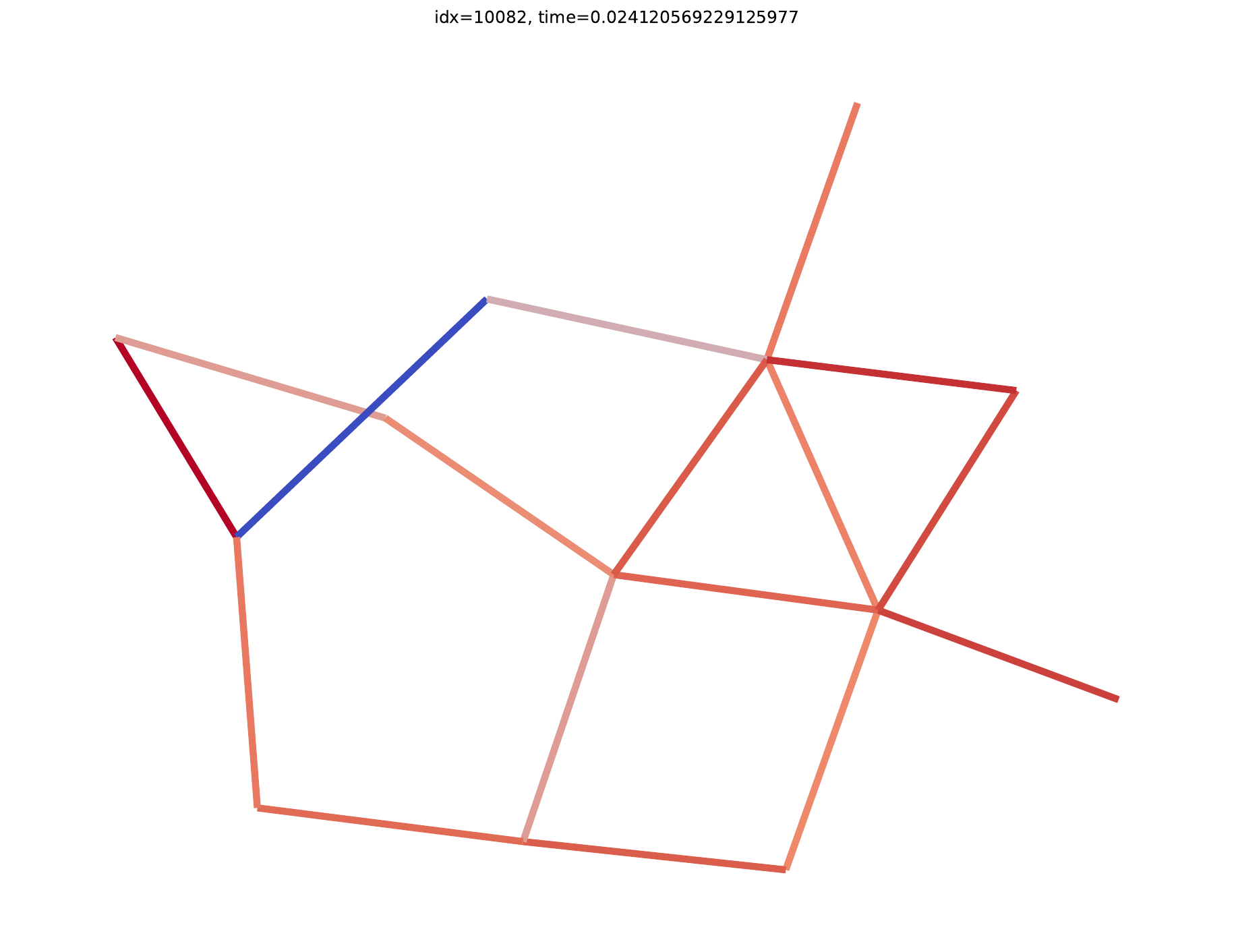} &
\imgcell{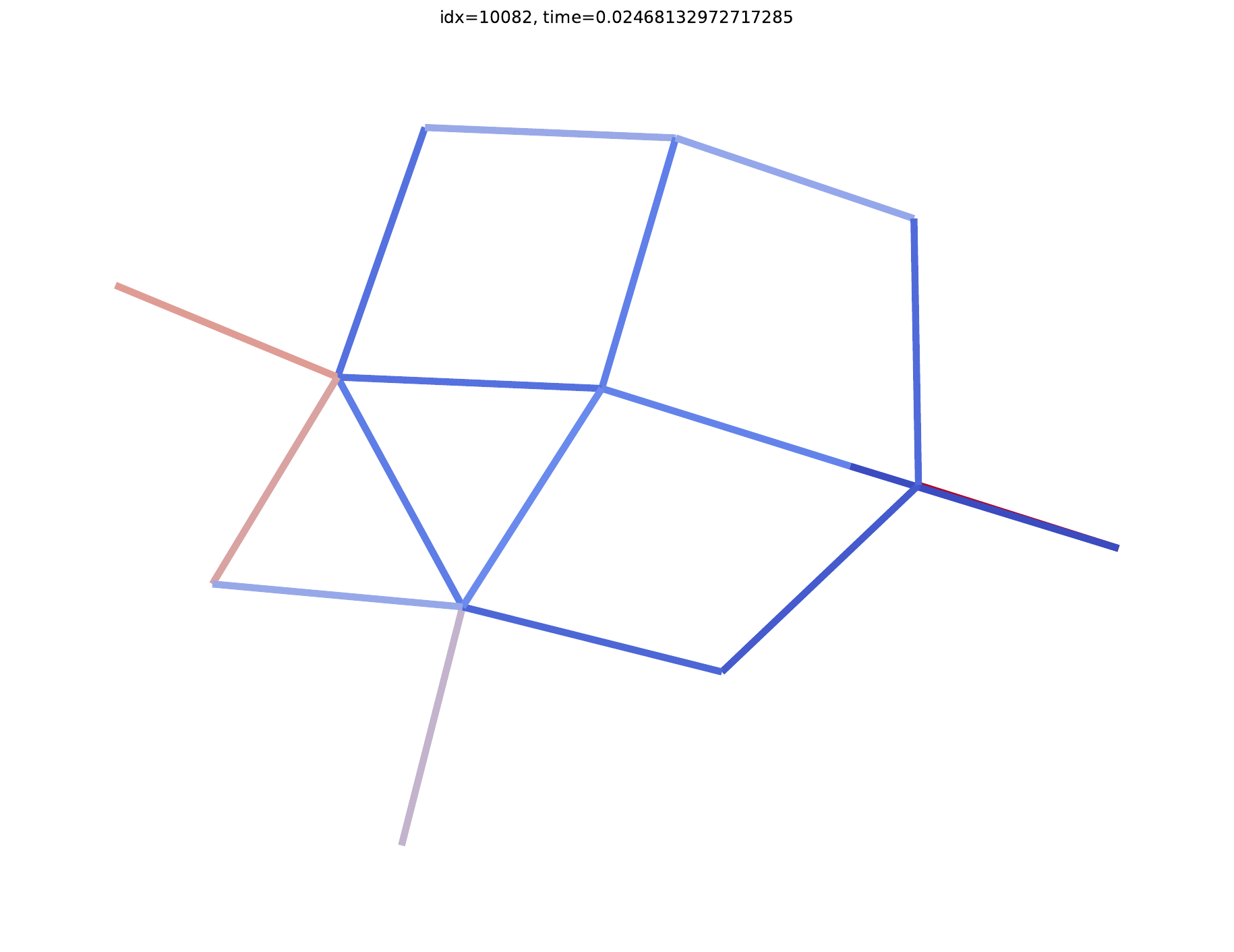} \\

&
t = 0.00s &
t = 0.22s &
t = 0.02s &
t = 0.04s &
t = 96.73s &
t = 0.02s &
t = 0.02s &
t = 0.03s &
t = 0.02s &
t = 0.03s &
t = 0.02s &
t = 0.02s \\

\makecell{\bfseries grafo7185.55\\N = 61\\M = 88} &
\imgcell{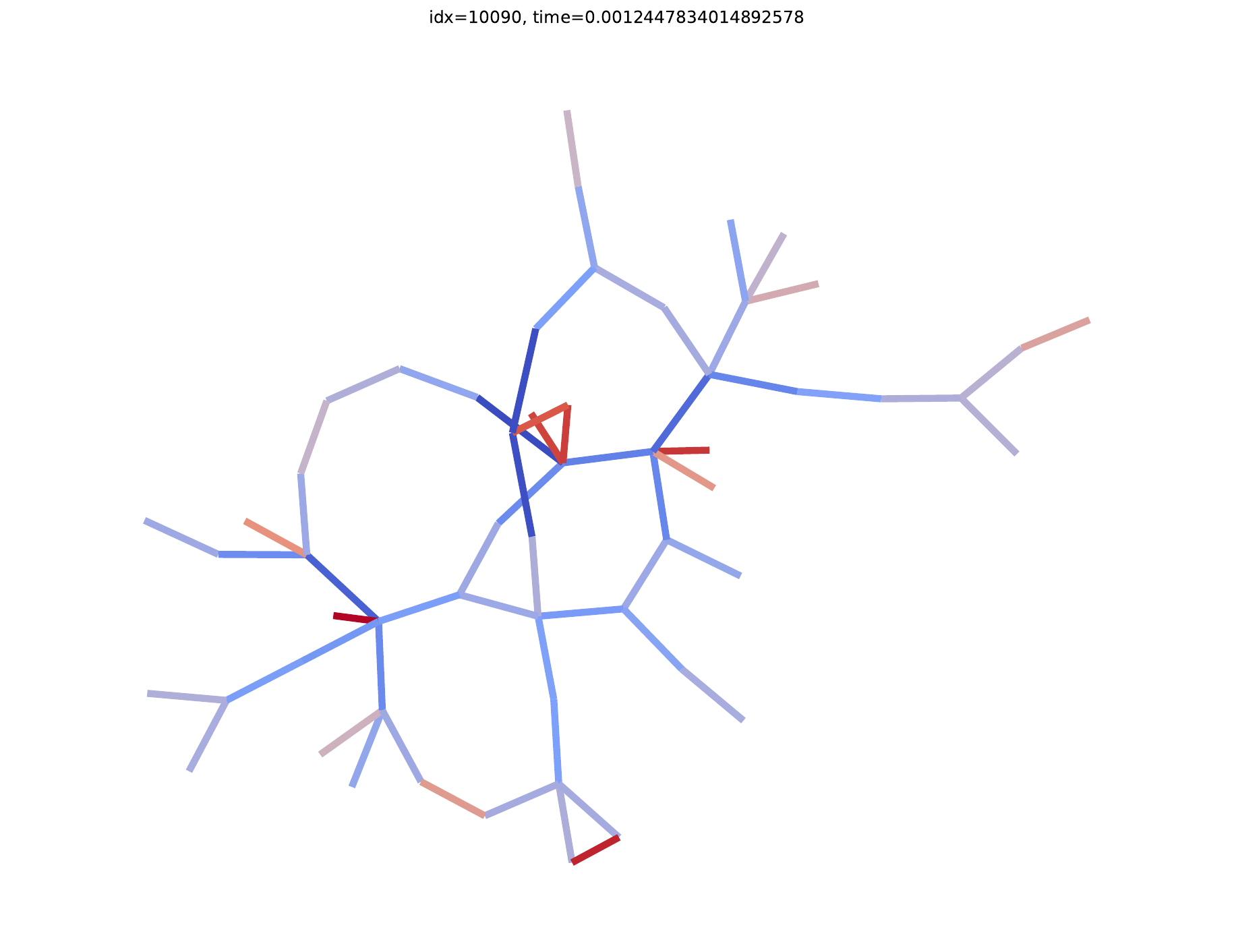} &
\imgcell{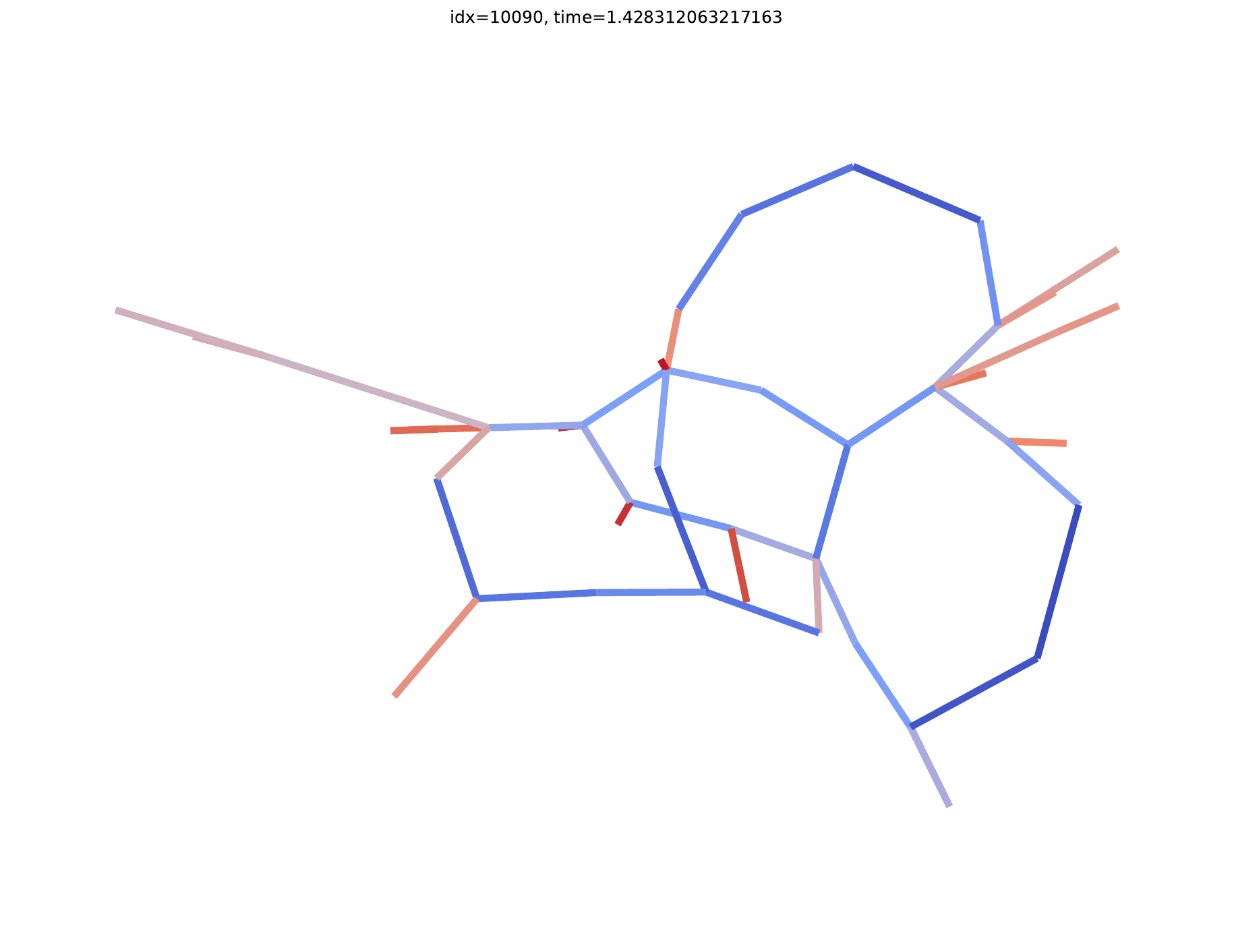} &
\imgcell{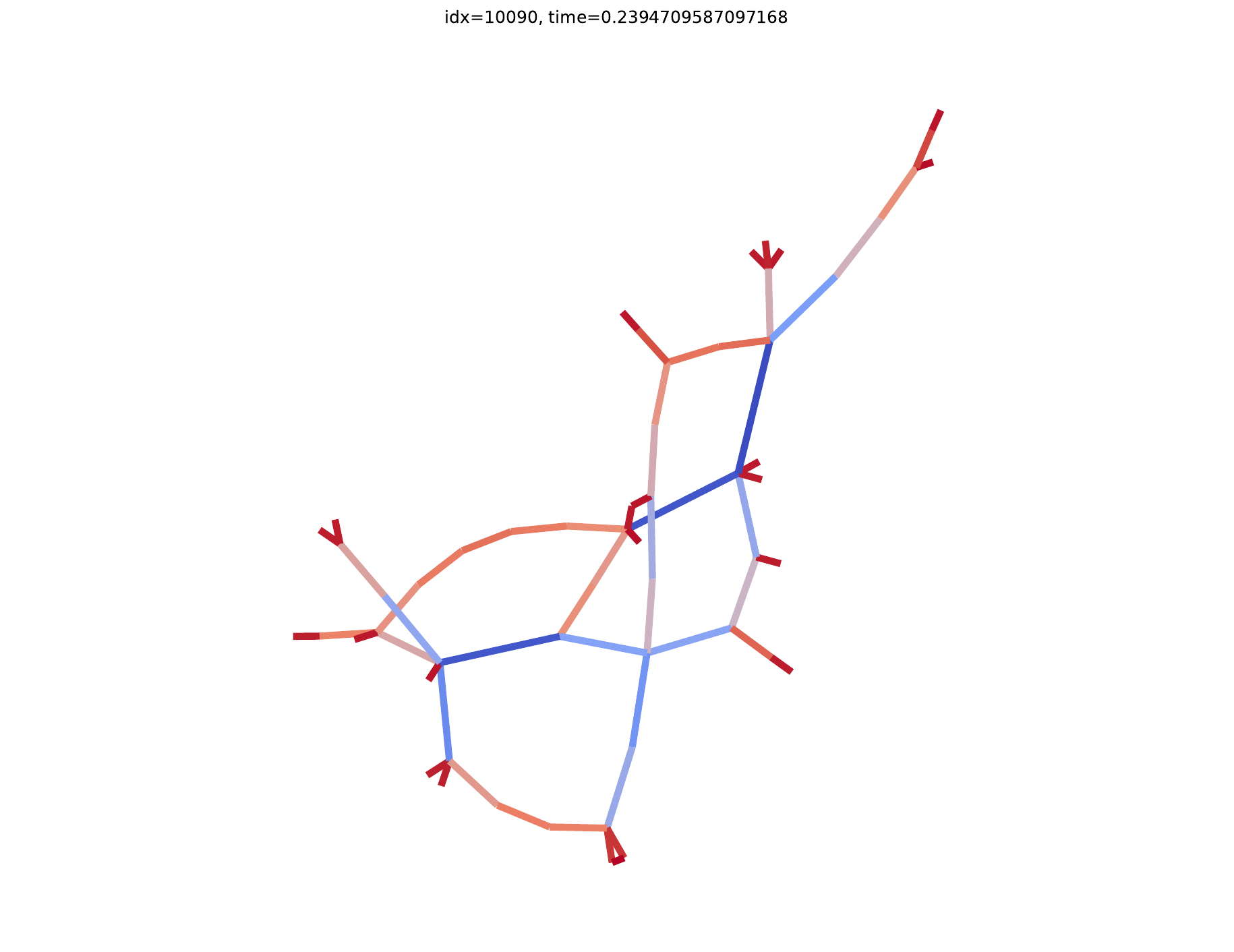} &
\imgcell{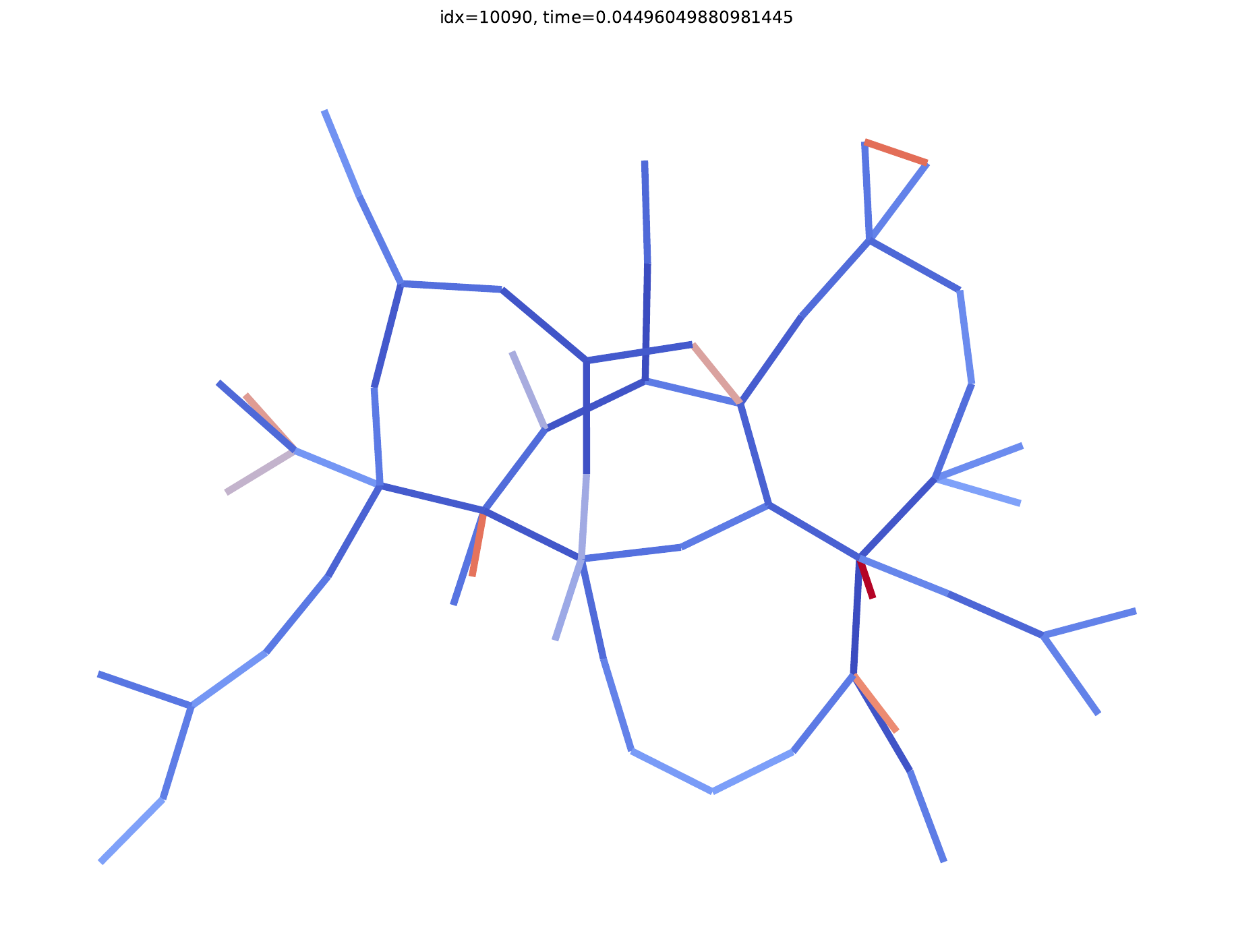} &
\imgcell{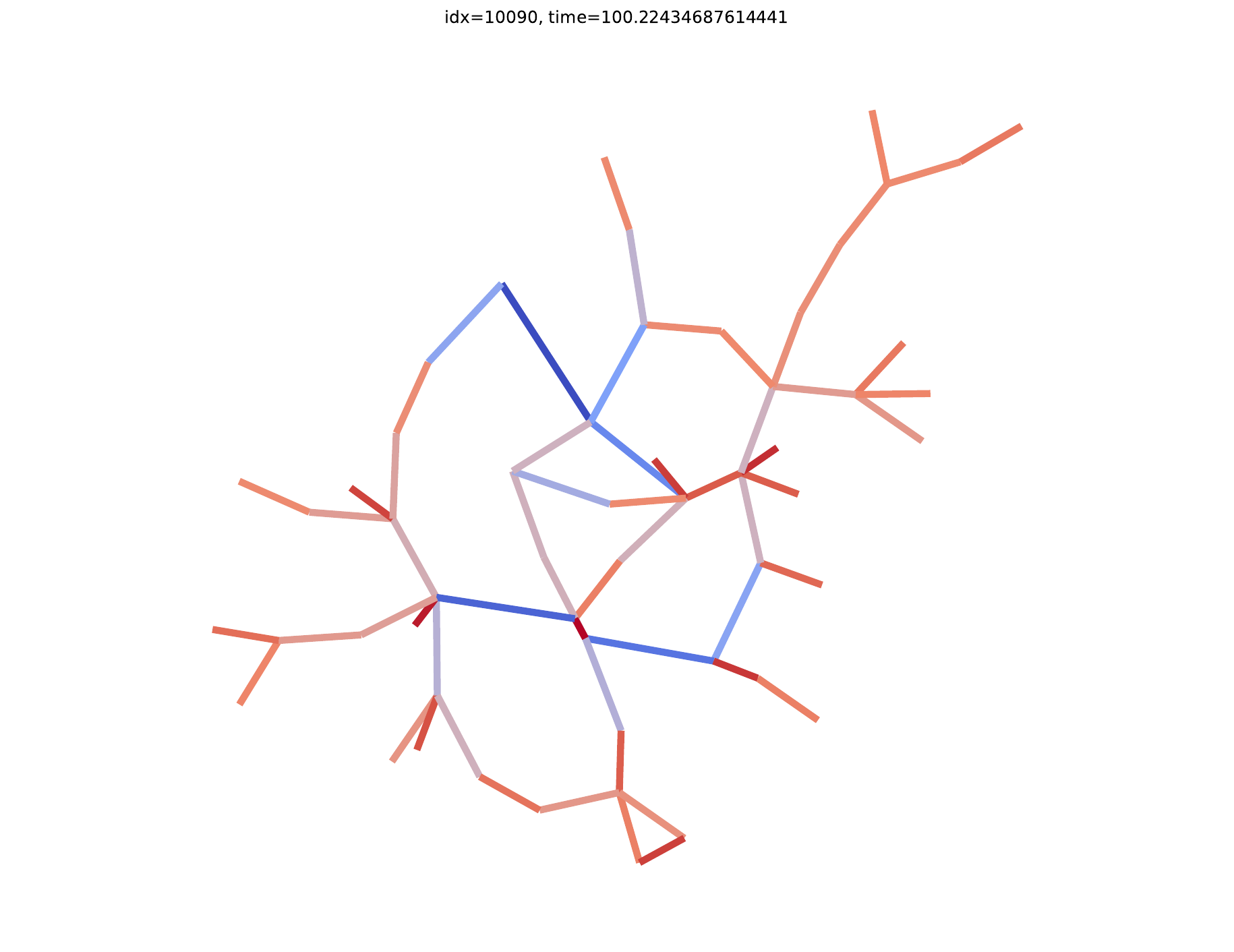} &
\imgcell{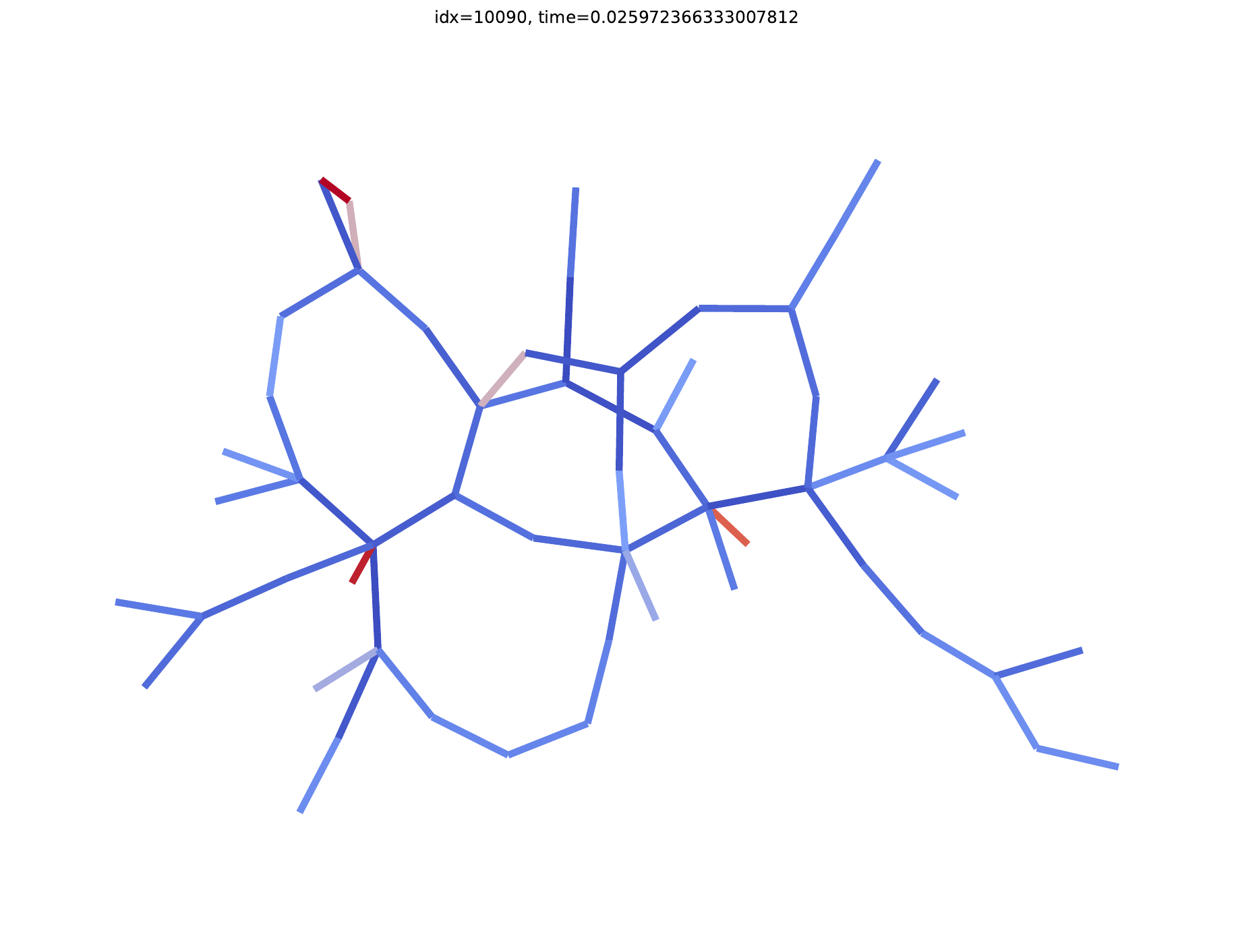} &
\imgcell{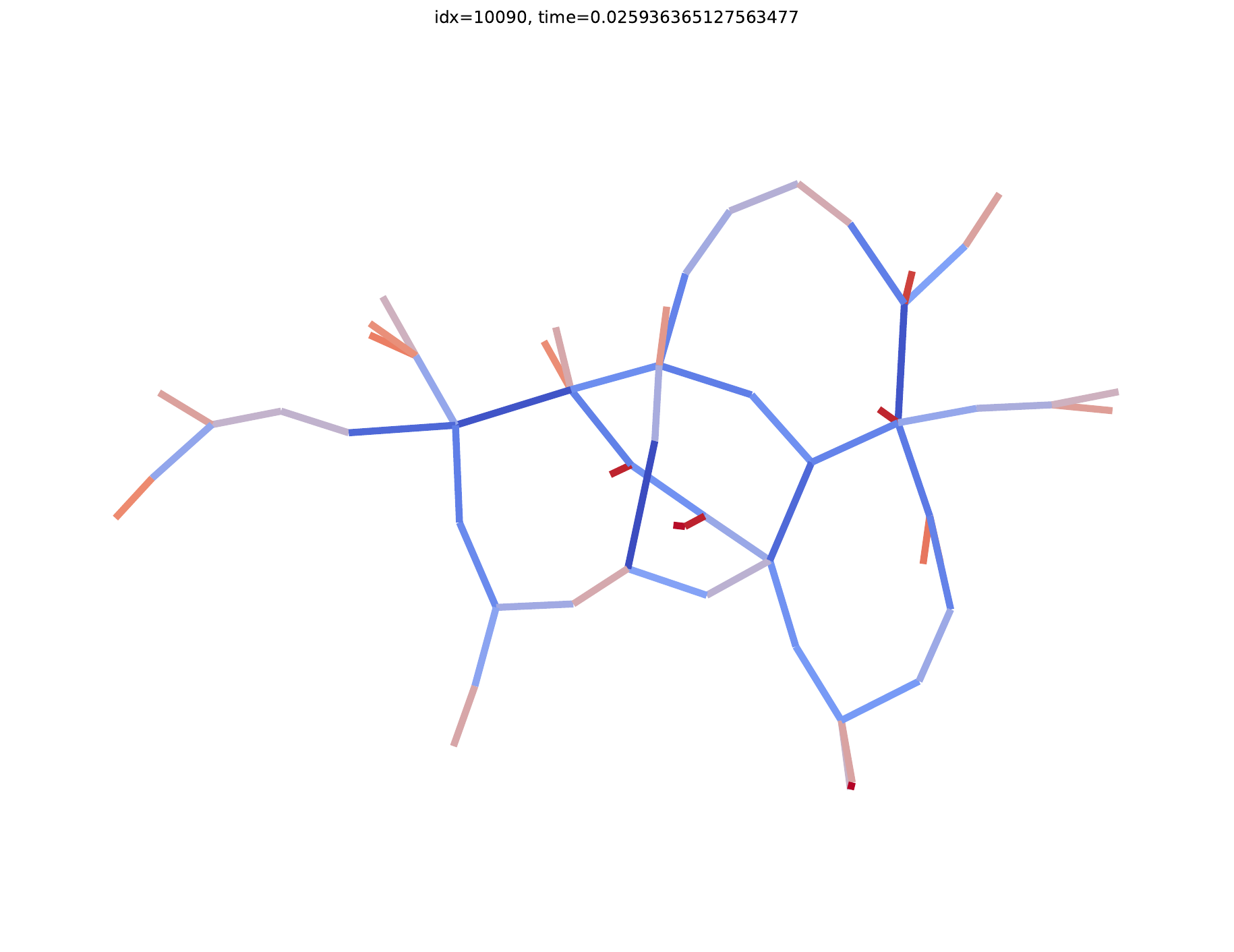} &
\imgcell{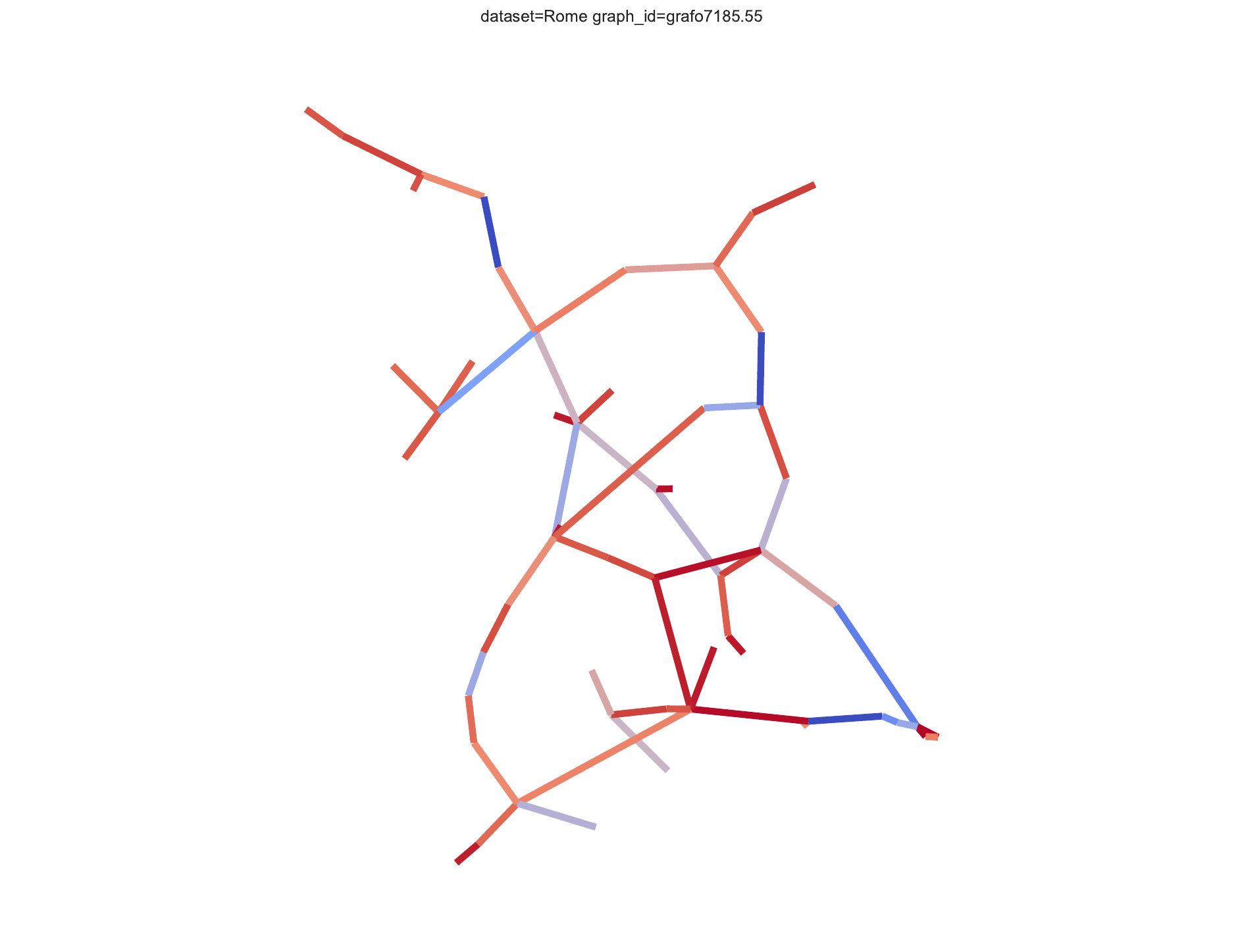} &
\imgcell{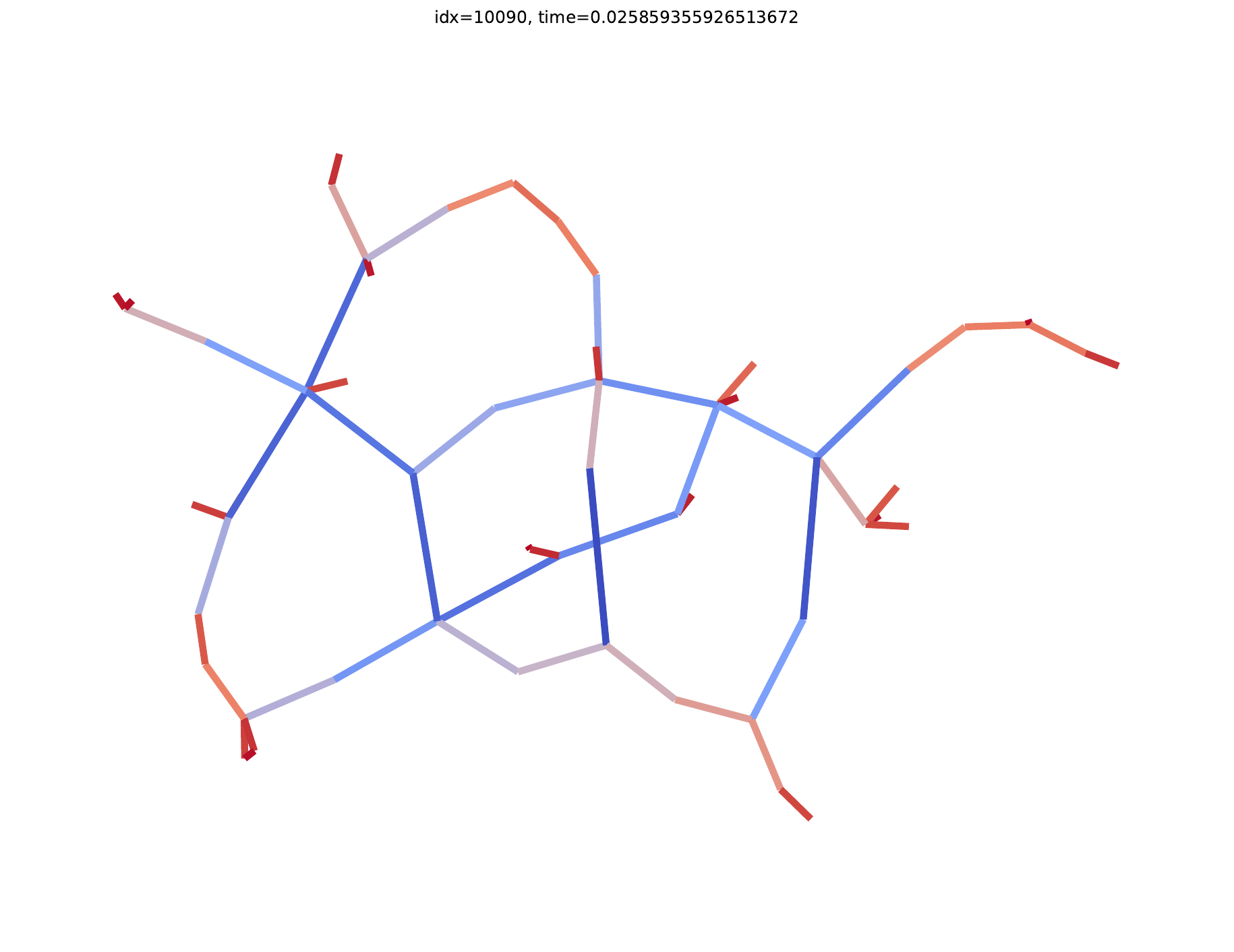} &
\imgcell{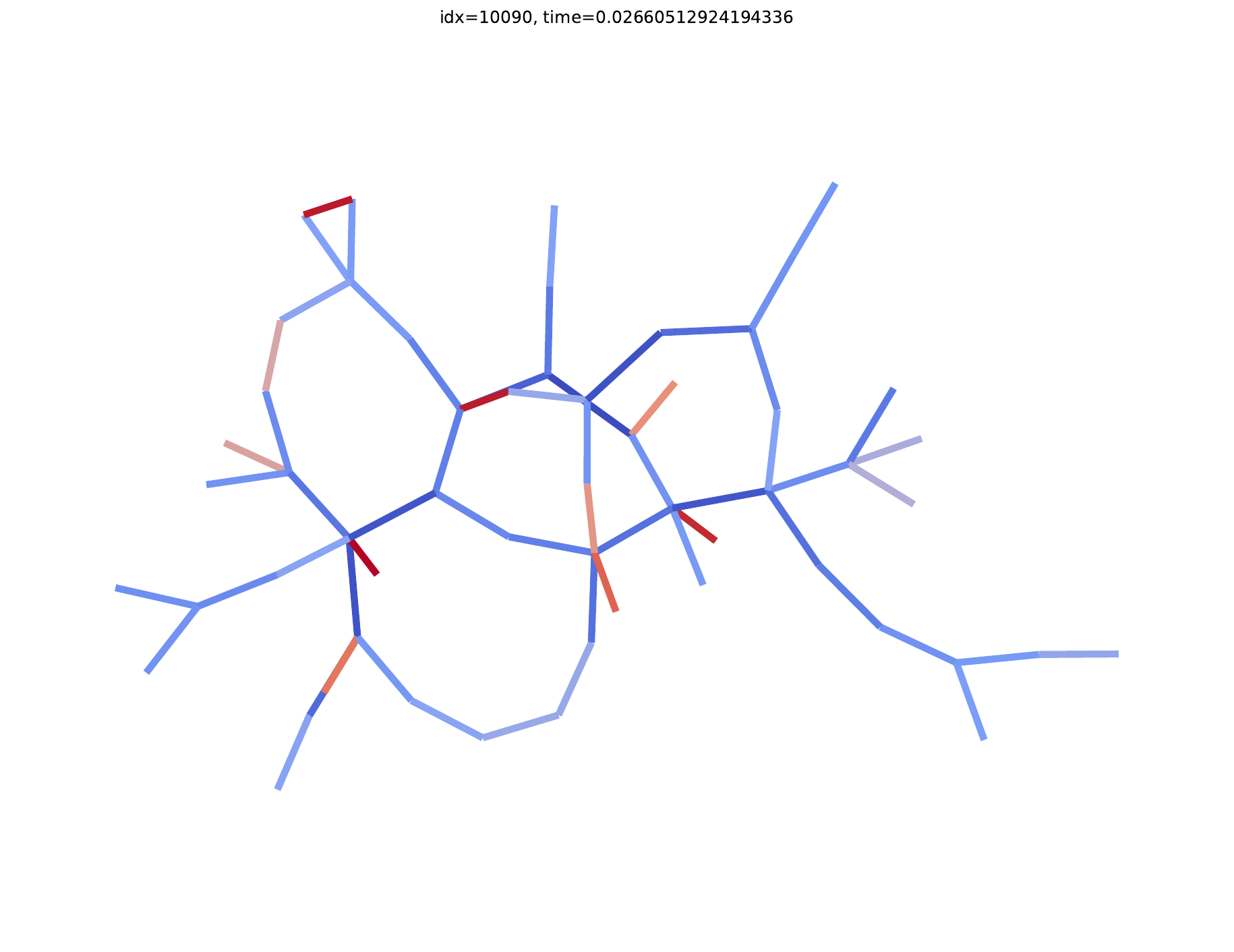} &
\imgcell{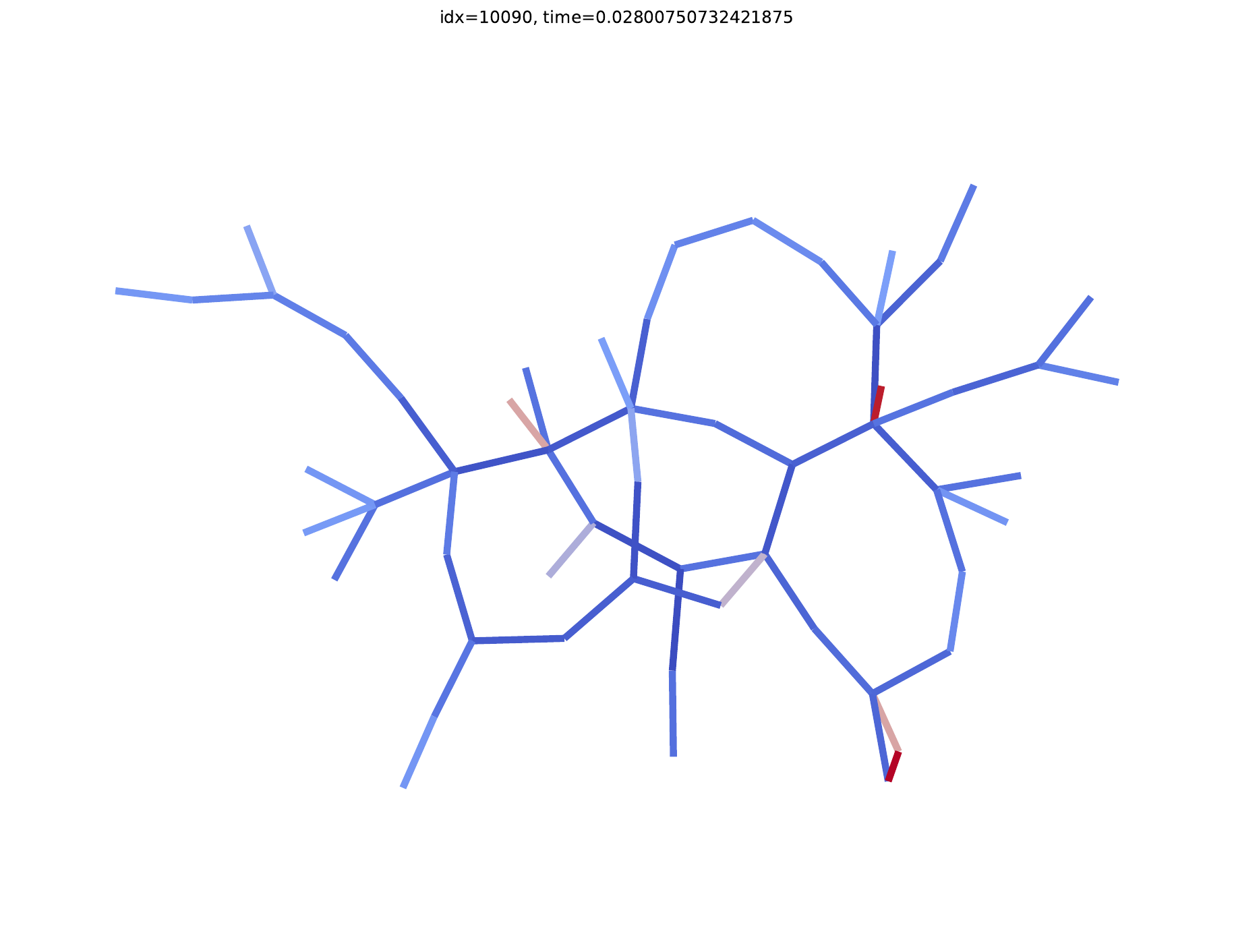} &
\imgcell{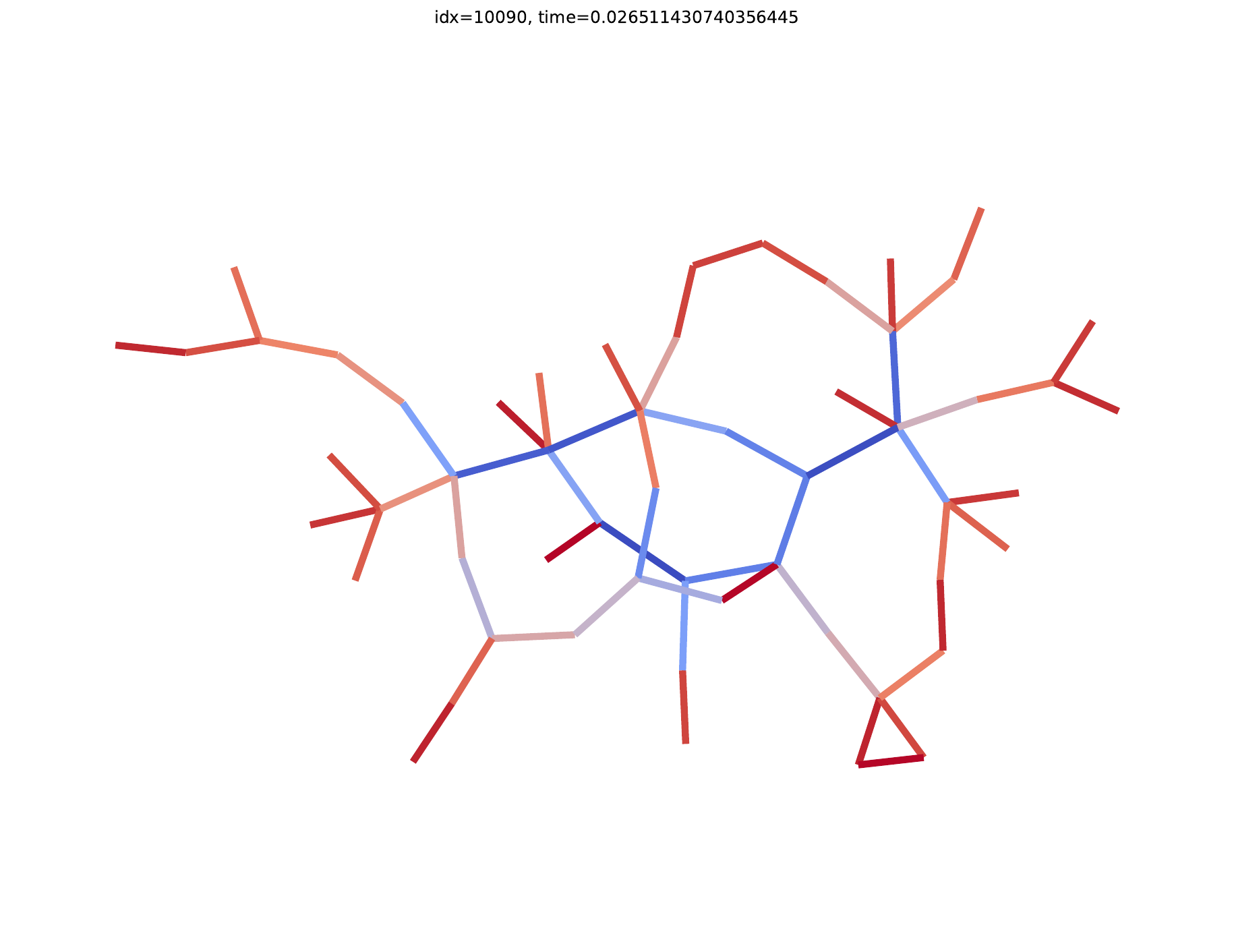} \\

&
t = 0.00s &
t = 1.43s &
t = 0.24s &
t = 0.04s &
t = 100.22s &
t = 0.03s &
t = 0.03s &
t = 0.03s &
t = 0.03s &
t = 0.03s &
t = 0.03s &
t = 0.03s \\

\makecell{\bfseries grafo682.16\\N = 43\\M = 55} &
\imgcell{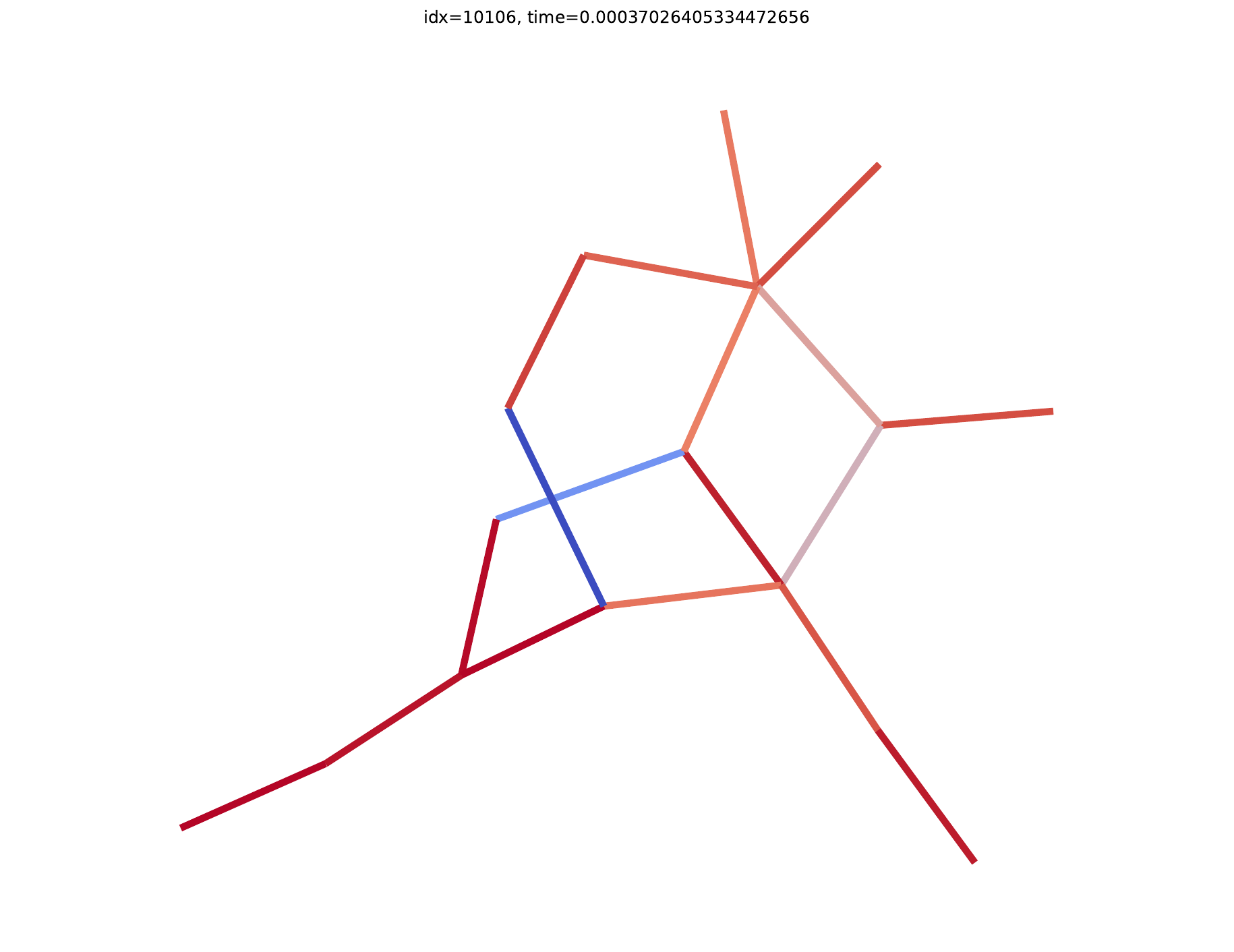} &
\imgcell{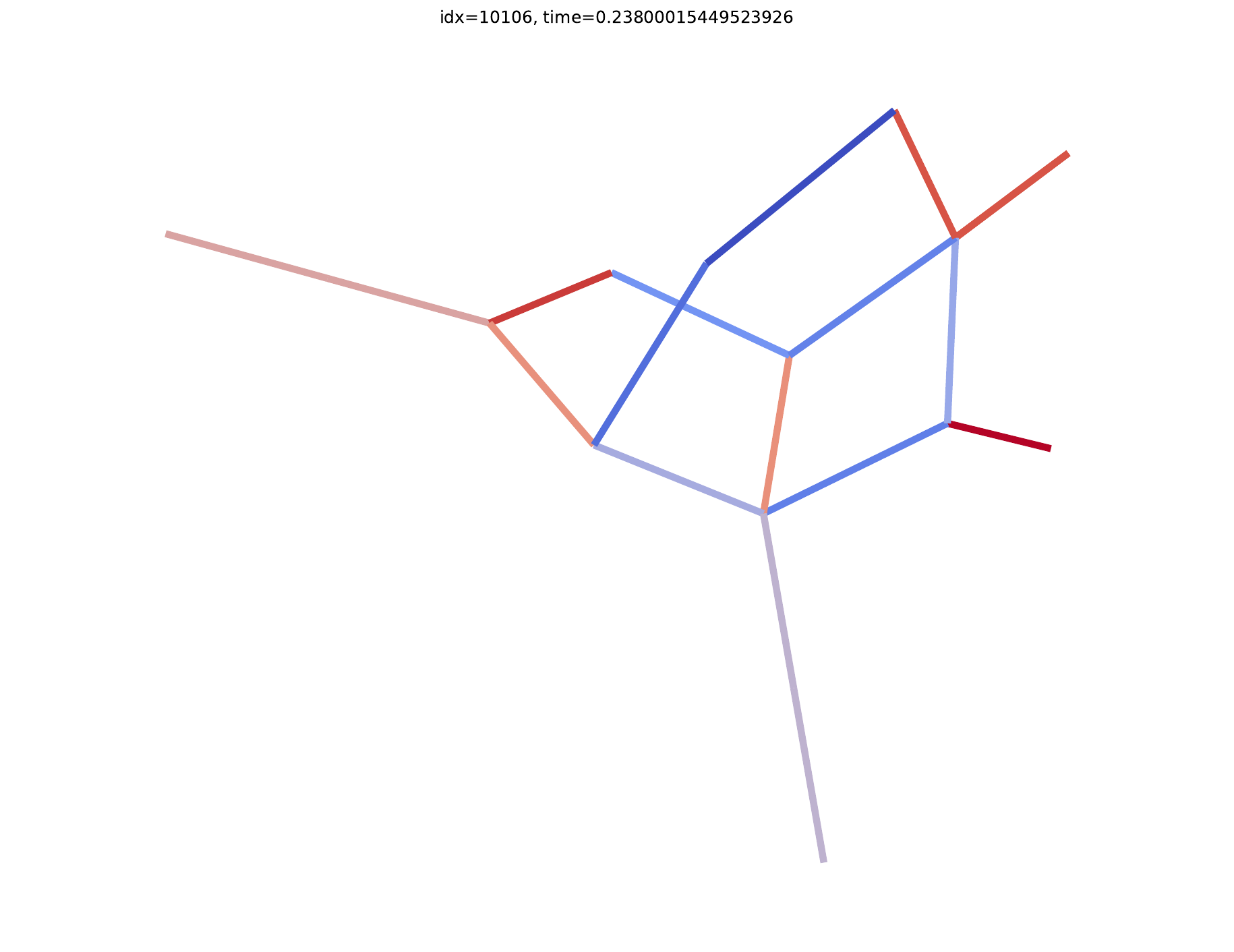} &
\imgcell{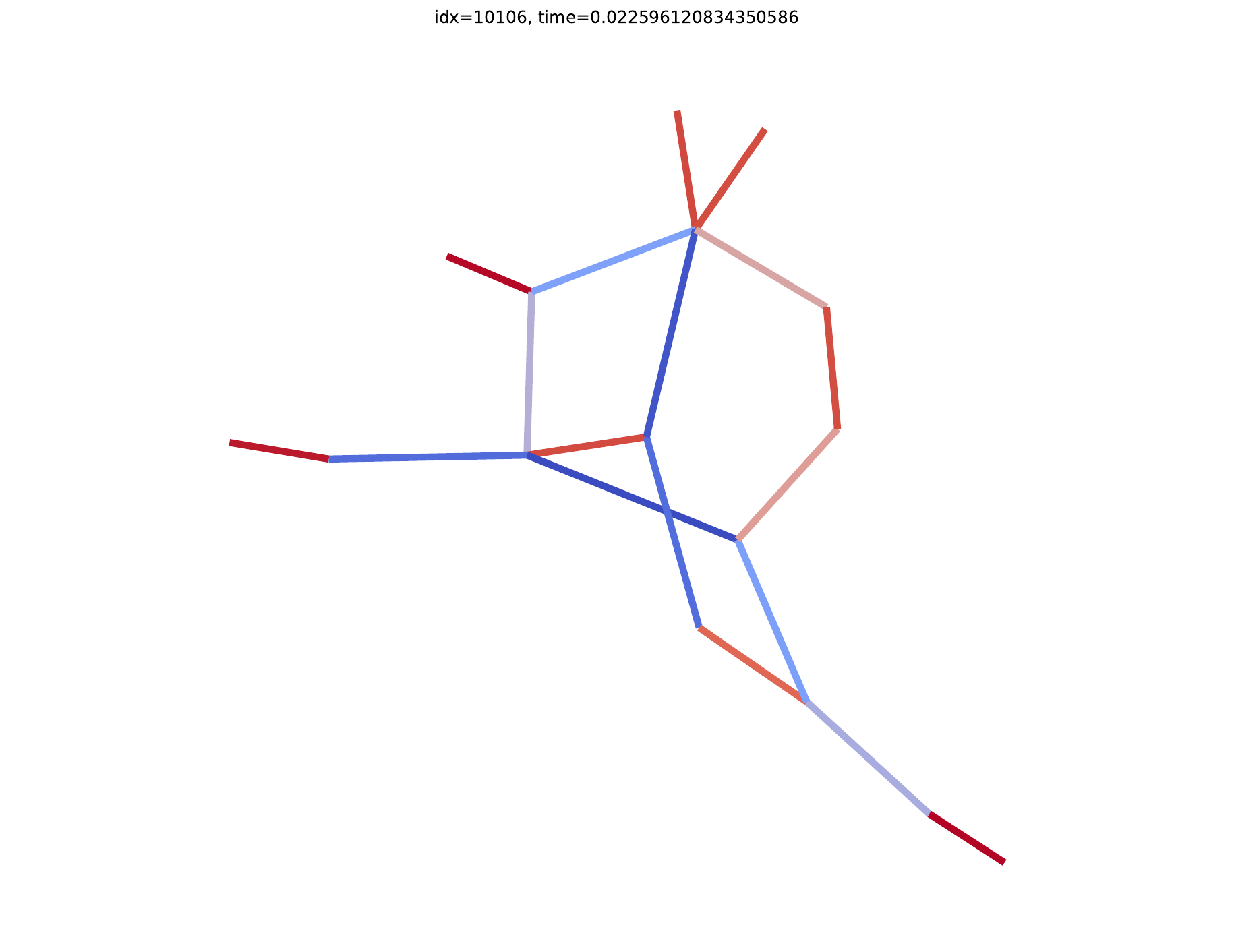} &
\imgcell{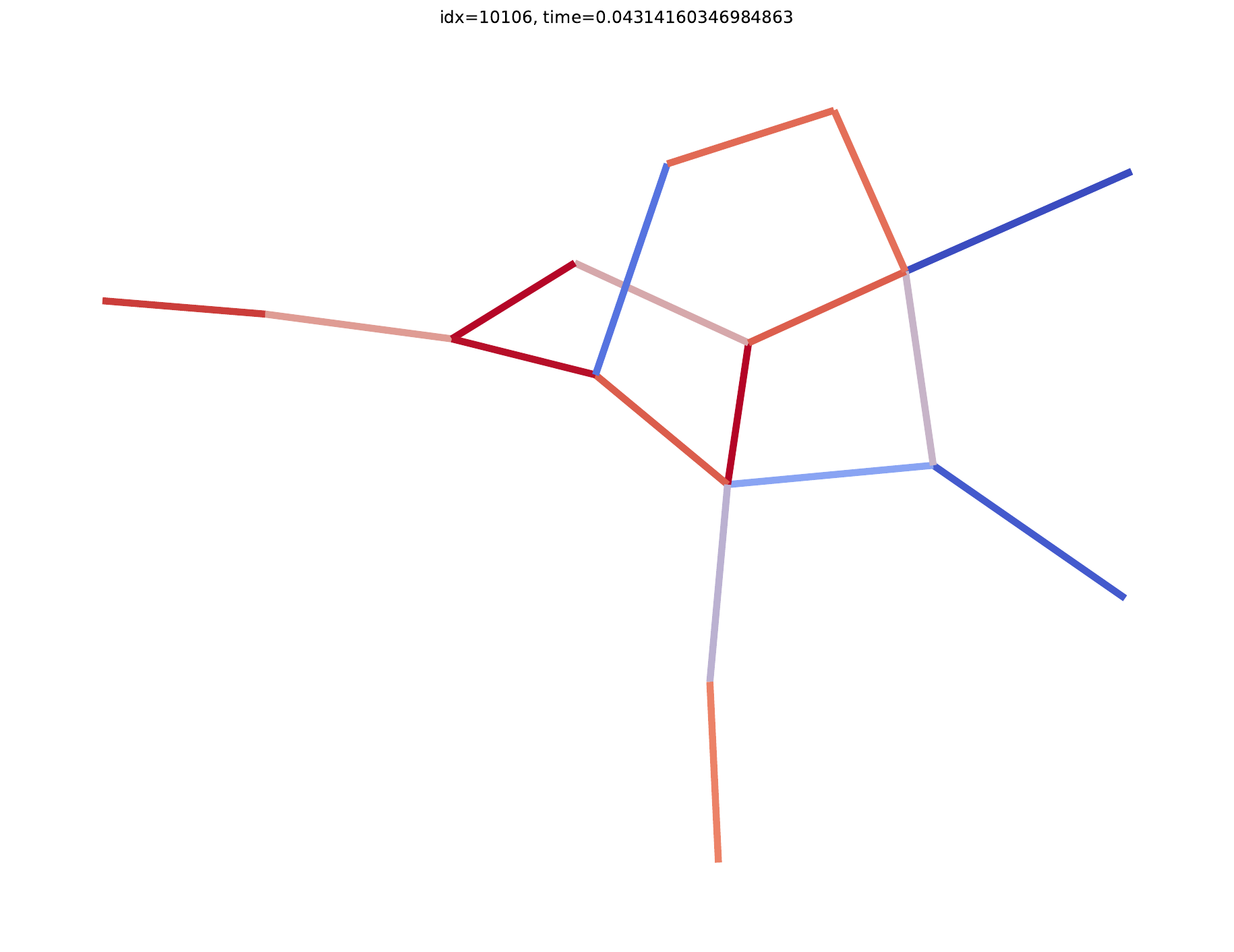} &
\imgcell{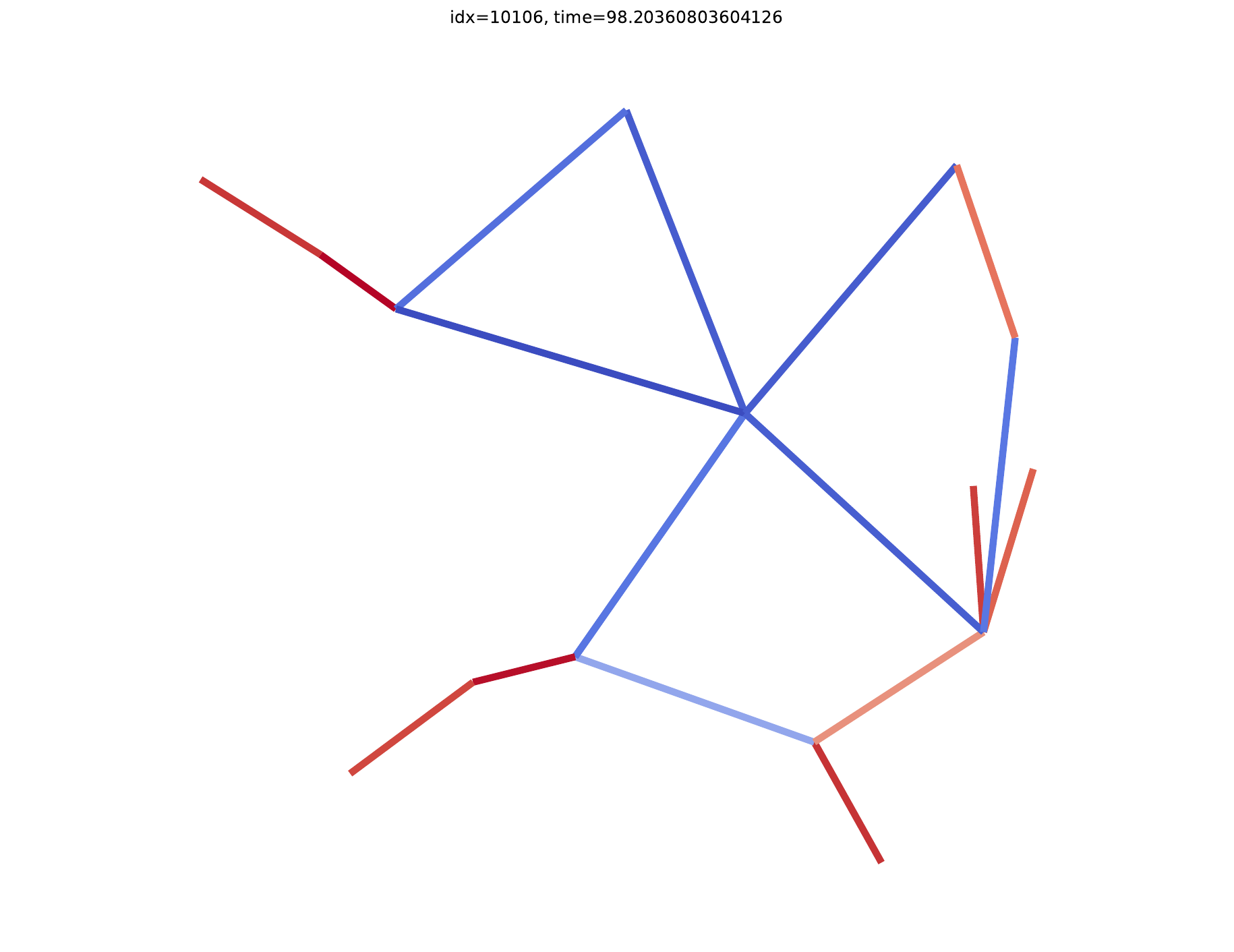} &
\imgcell{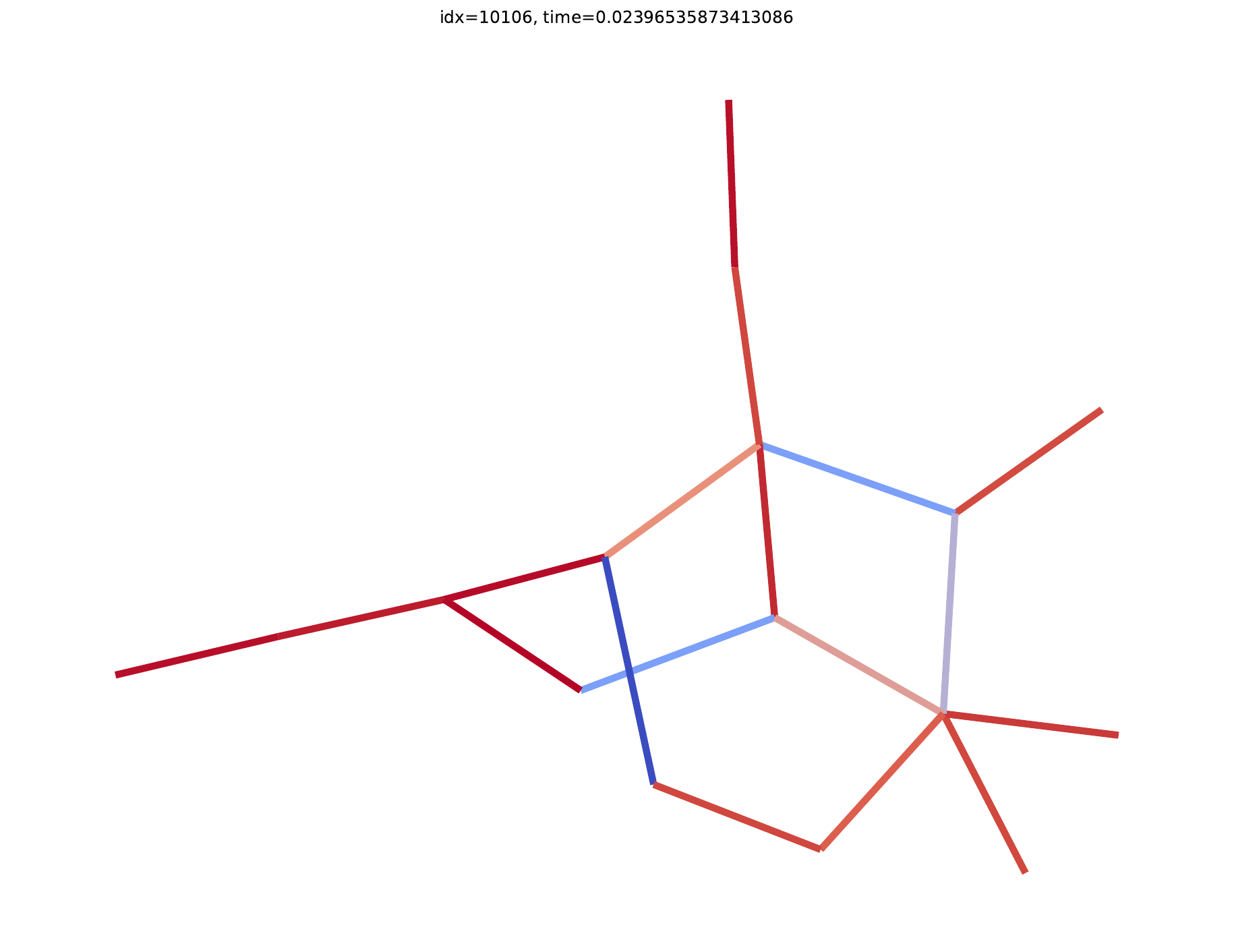} &
\imgcell{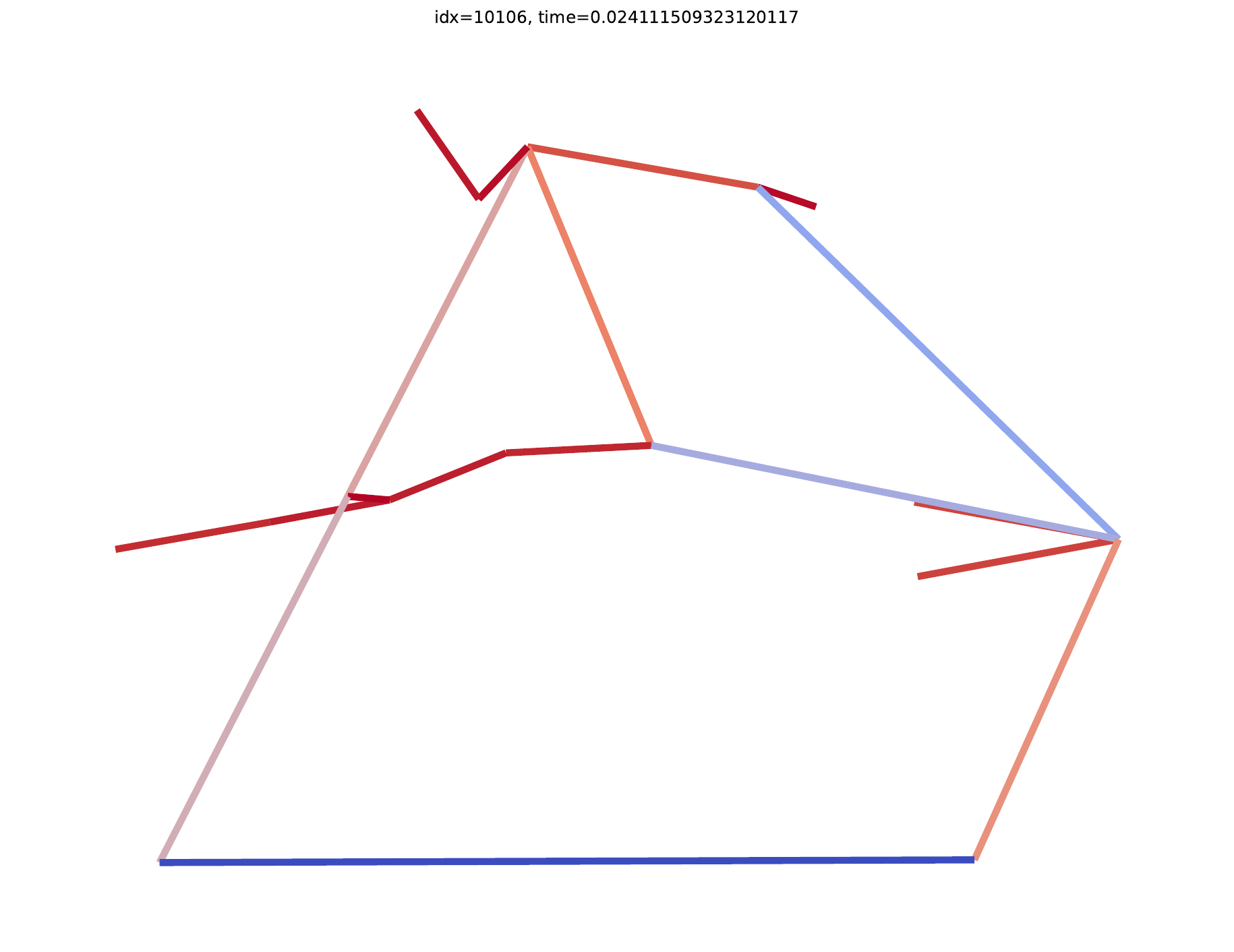} &
\imgcell{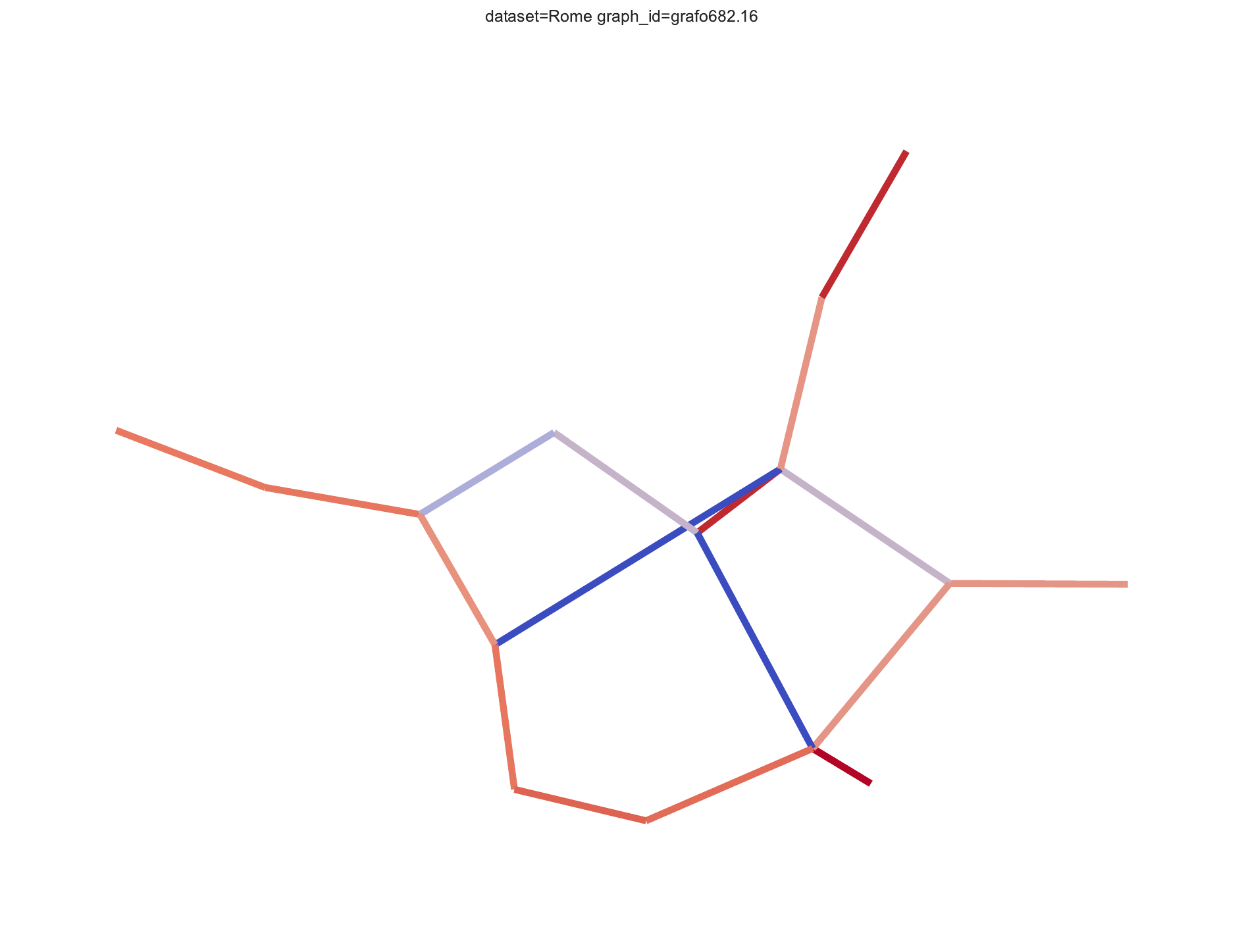} &
\imgcell{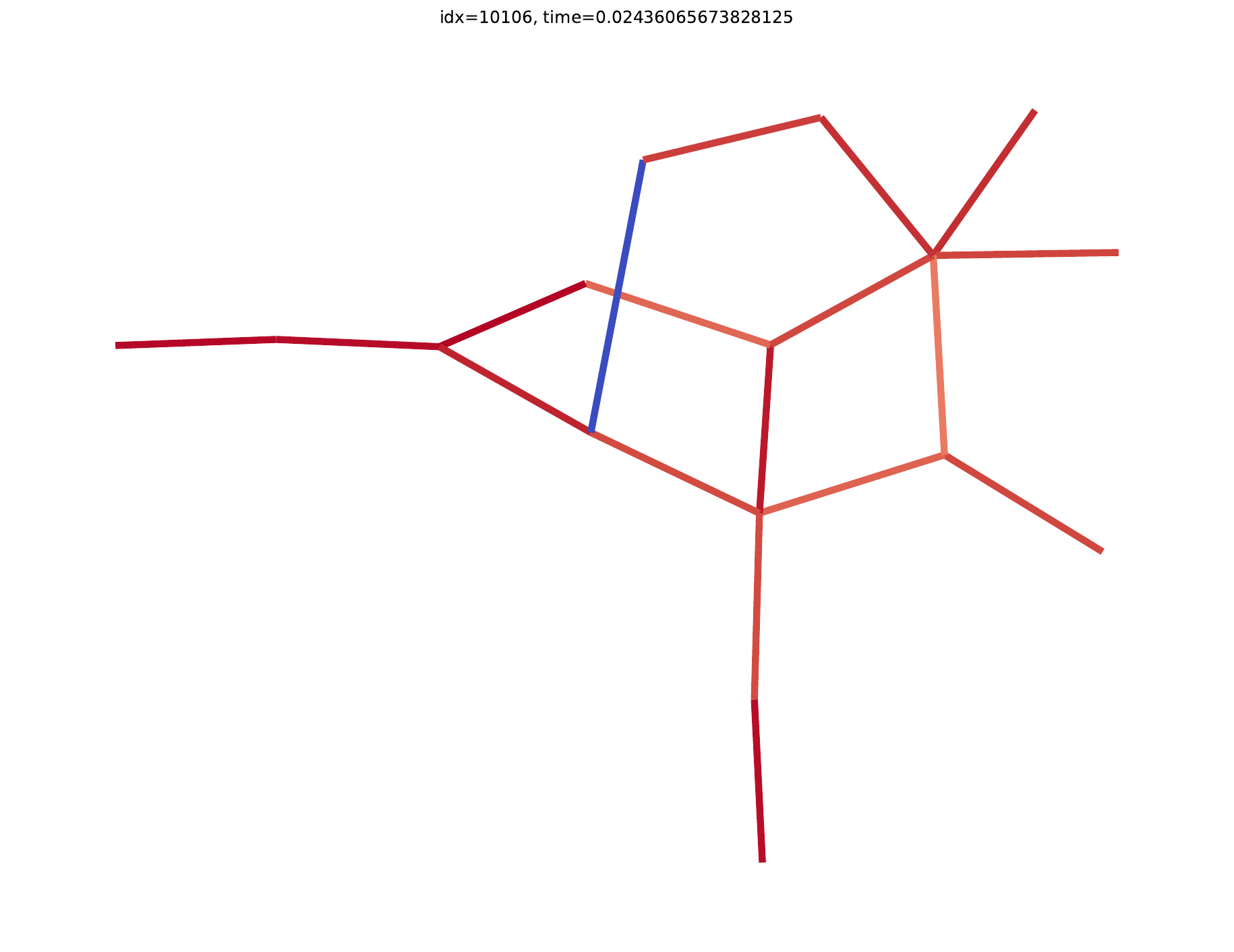} &
\imgcell{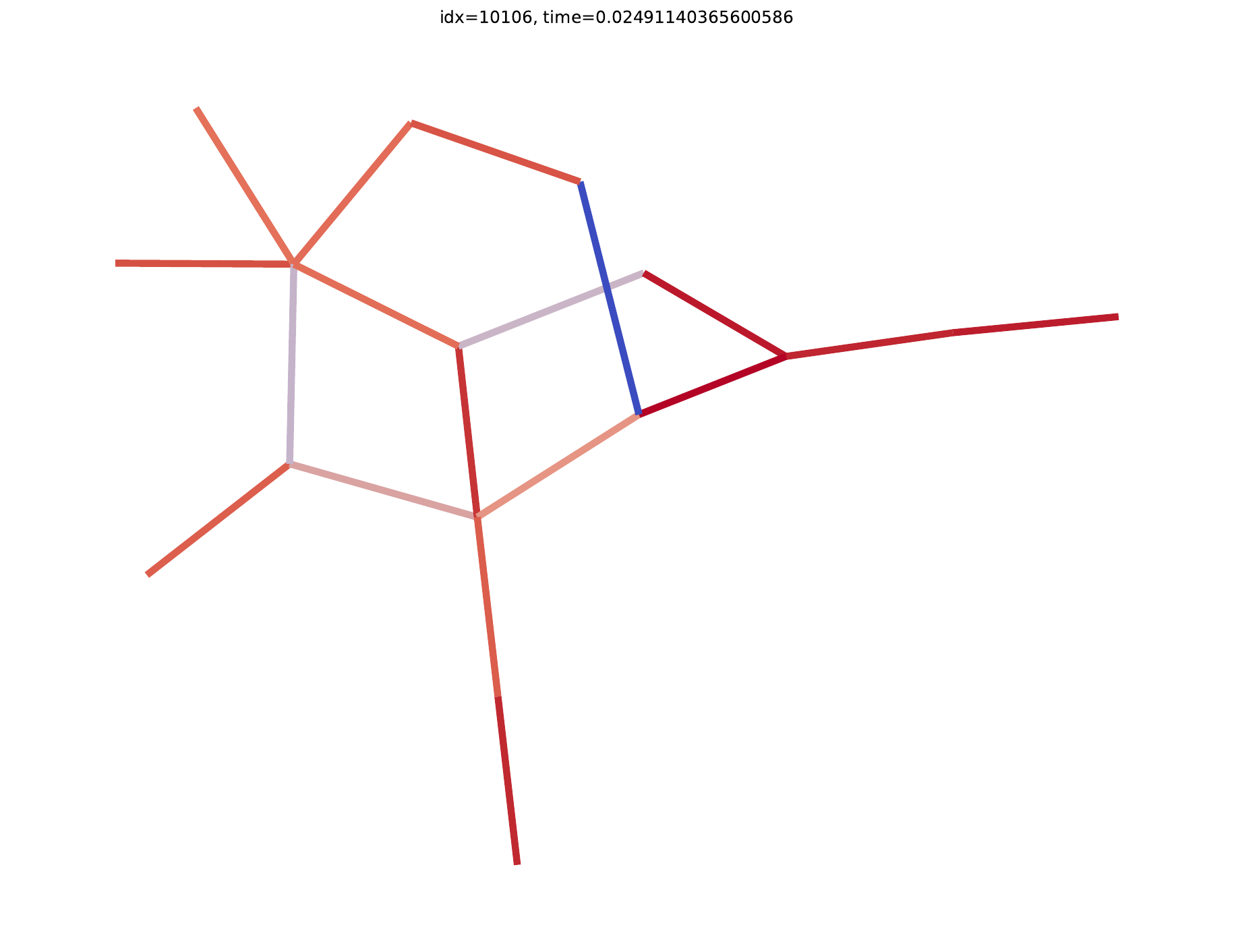} &
\imgcell{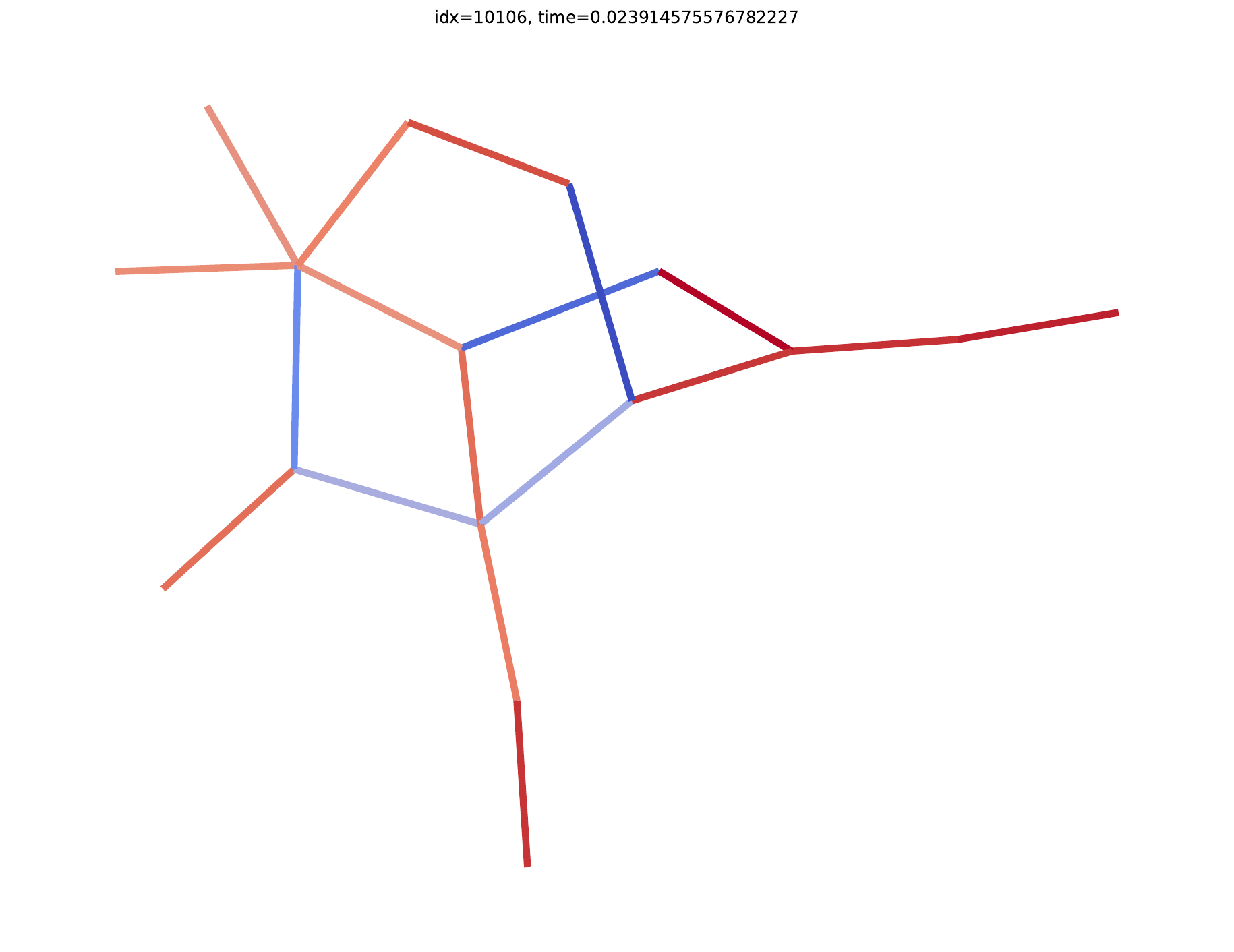} &
\imgcell{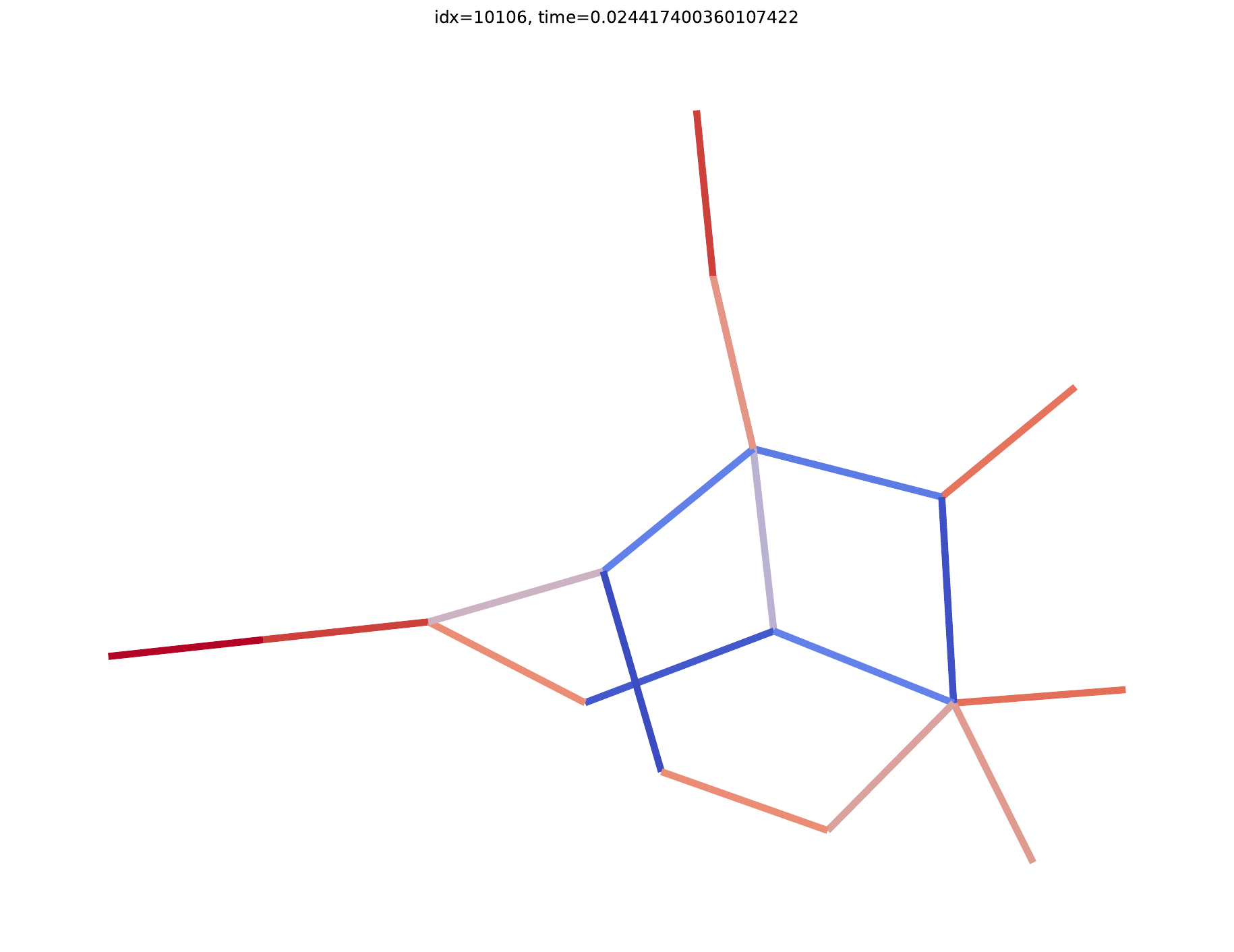} \\

&
t = 0.00s &
t = 0.24s &
t = 0.02s &
t = 0.04s &
t = 98.20s &
t = 0.02s &
t = 0.02s &
t = 0.02s &
t = 0.02s &
t = 0.02s &
t = 0.02s &
t = 0.02s \\

\makecell{\bfseries grafo8900.94\\N = 35\\M = 47} &
\imgcell{figures/rome_graphs/10111_sgd2.pdf} &
\imgcell{figures/rome_graphs/10111_pmds.pdf} &
\imgcell{figures/rome_graphs/10111_fa2.pdf} &
\imgcell{figures/rome_graphs/10111_deepgd.pdf} &
\imgcell{figures/rome_graphs/10111_gd2_stress_xing.pdf} &
\imgcell{figures/rome_graphs/10111_smartgd_stress.pdf} &
\imgcell{figures/rome_graphs/10111_smartgd_xing.pdf} &
\imgcell{figures/rome_graphs/10111_smartgd_xing_nsc.pdf} &
\imgcell{figures/rome_graphs/10111_smartgd_xangle.pdf} &
\imgcell{figures/rome_graphs/10111_smartgd_stress_xing.pdf} &
\imgcell{figures/rome_graphs/10111_smartgd_stress_xangle.pdf} &
\imgcell{figures/rome_graphs/10111_smartgd_combined.pdf} \\

&
t = 0.00s &
t = 0.78s &
t = 0.61s &
t = 0.05s &
t = 102.27s &
t = 0.04s &
t = 0.04s &
t = 0.03s &
t = 0.05s &
t = 0.04s &
t = 0.04s &
t = 0.05s \\

\makecell{\bfseries grafo2054.29\\N = 35\\M = 41} &
\imgcell{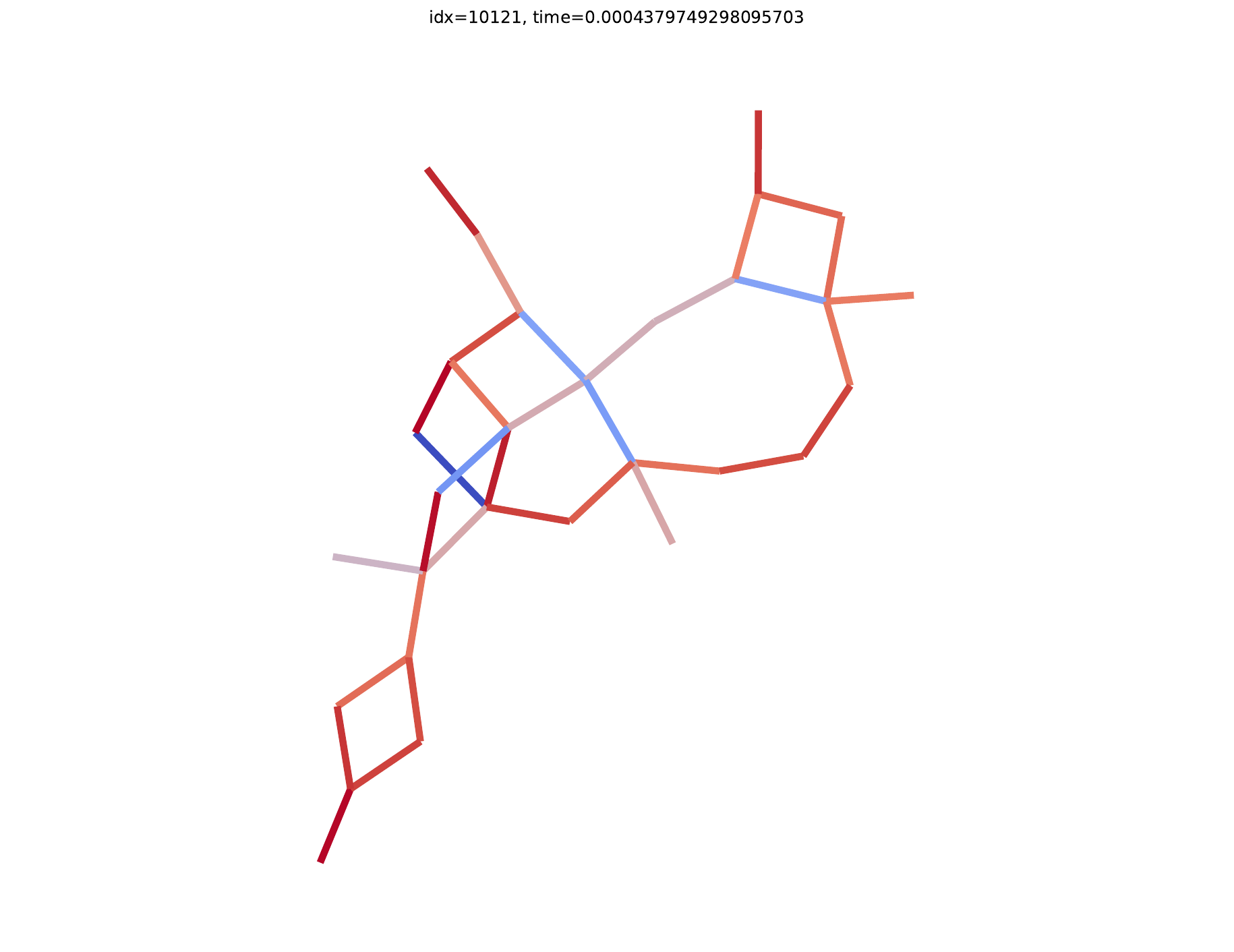} &
\imgcell{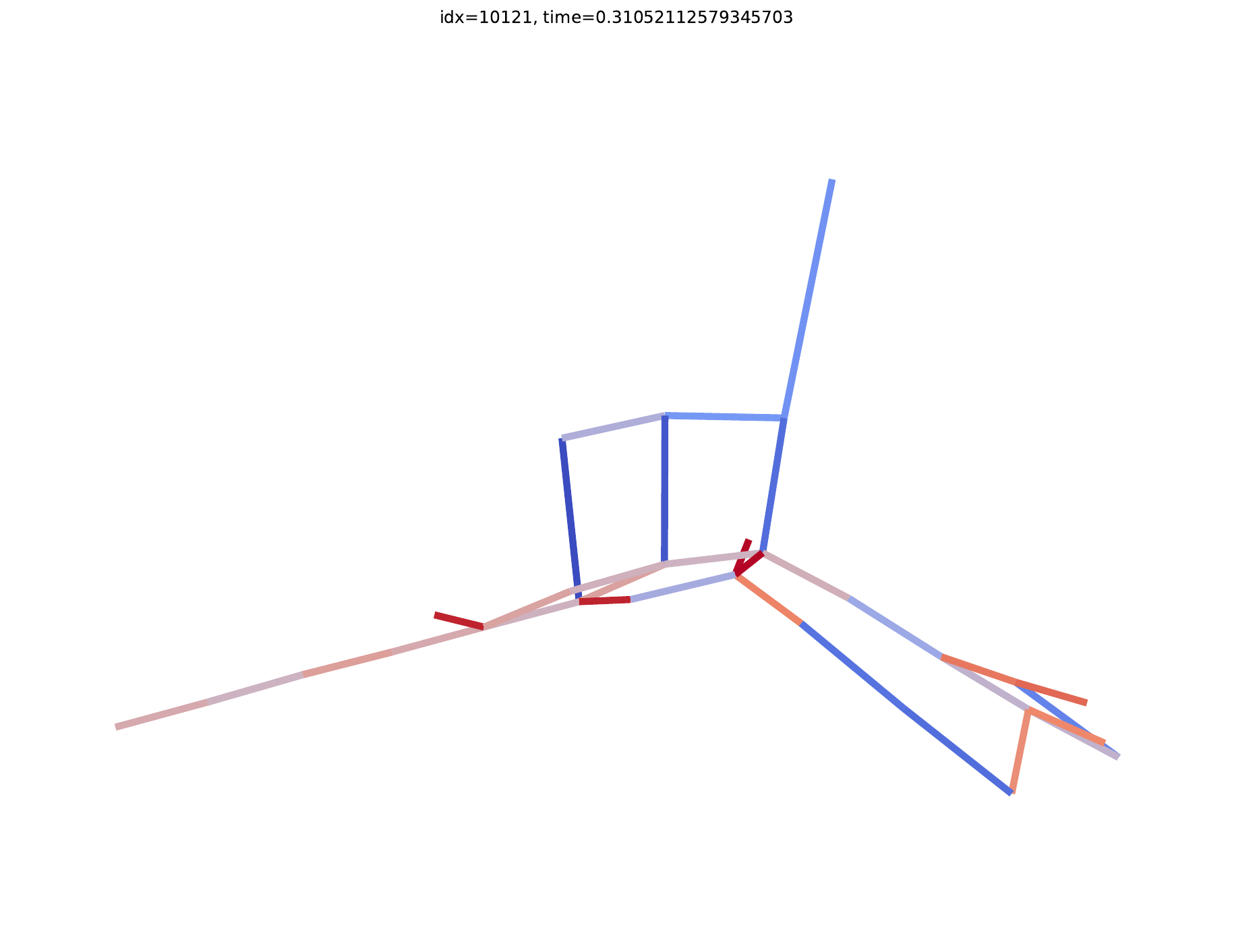} &
\imgcell{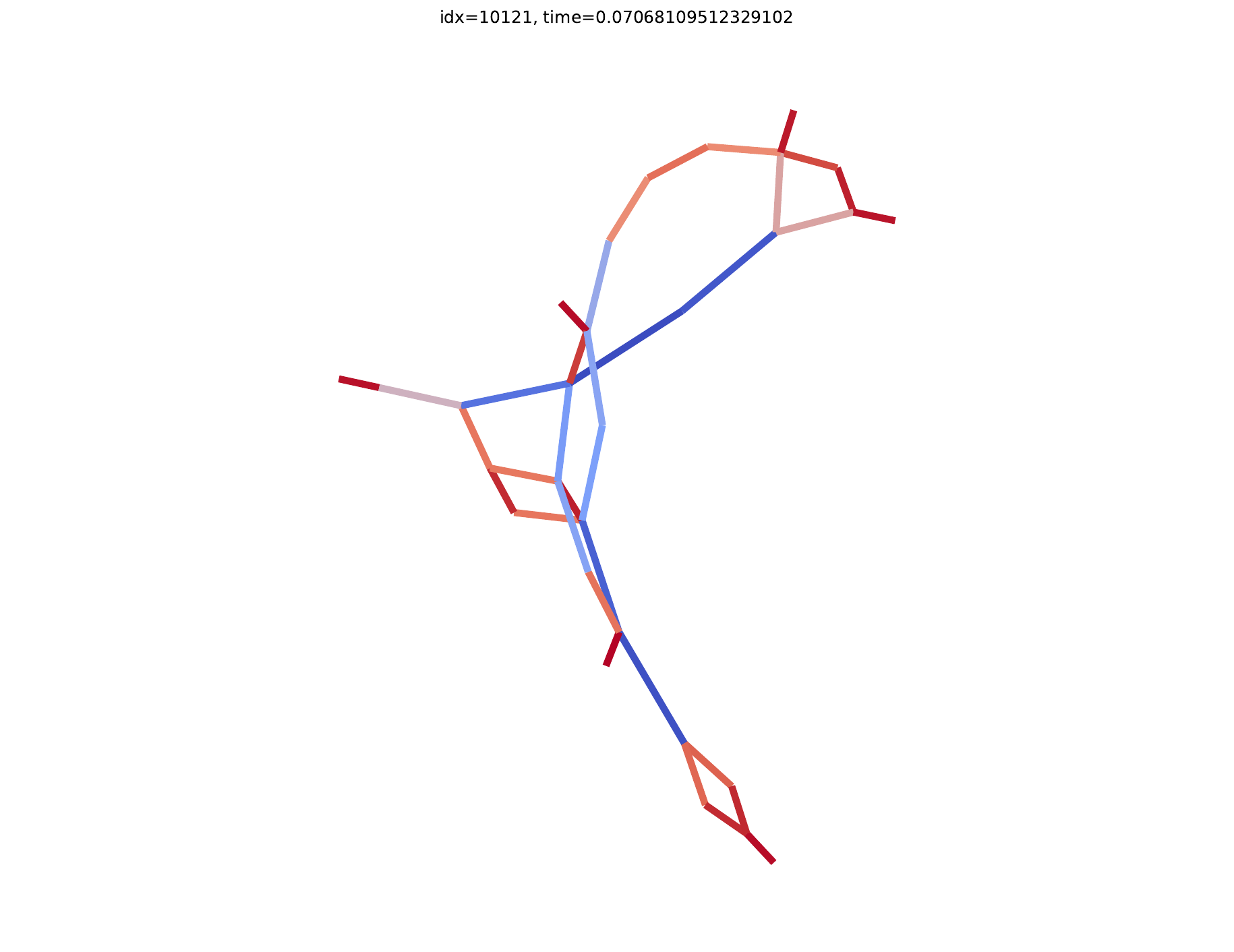} &
\imgcell{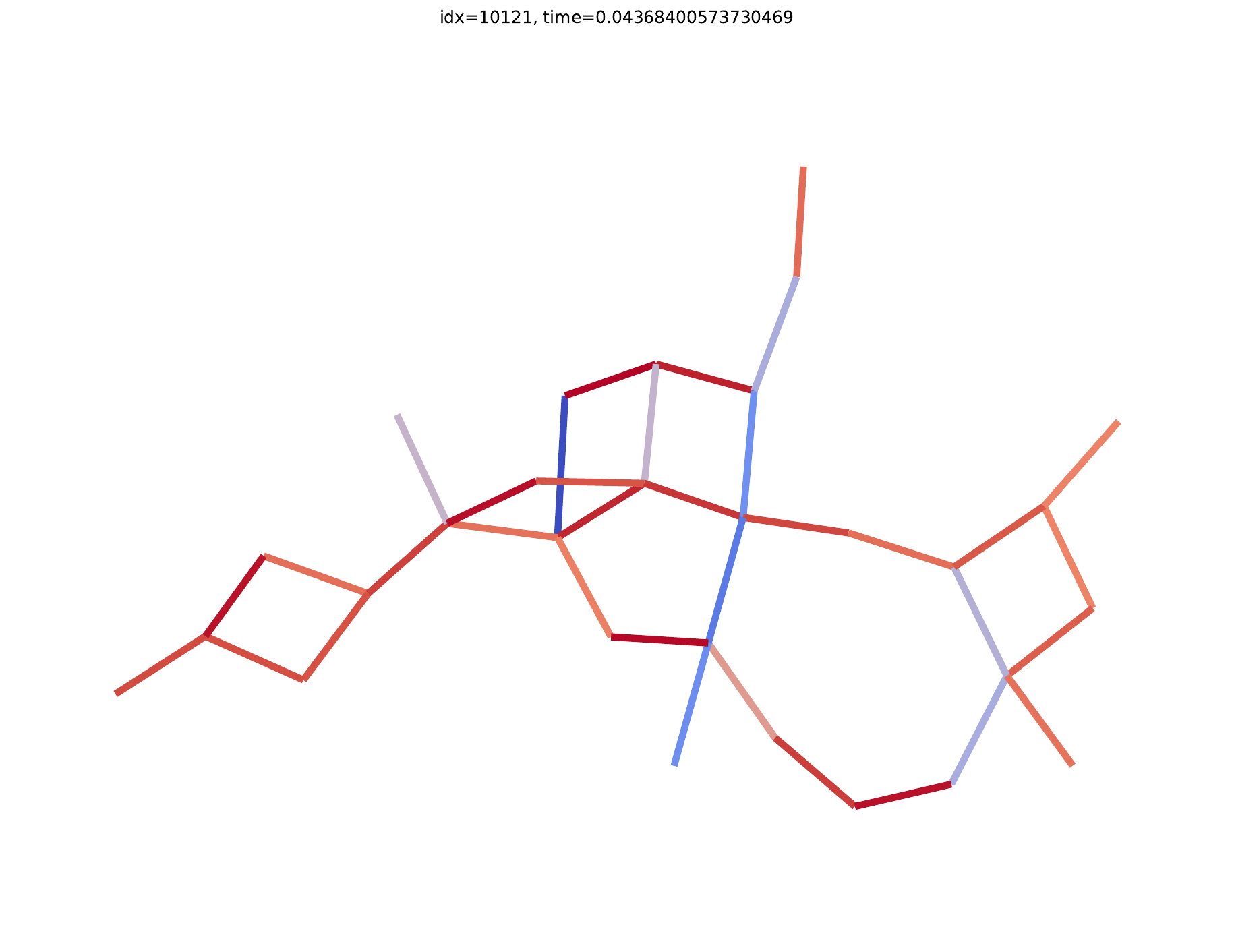} &
\imgcell{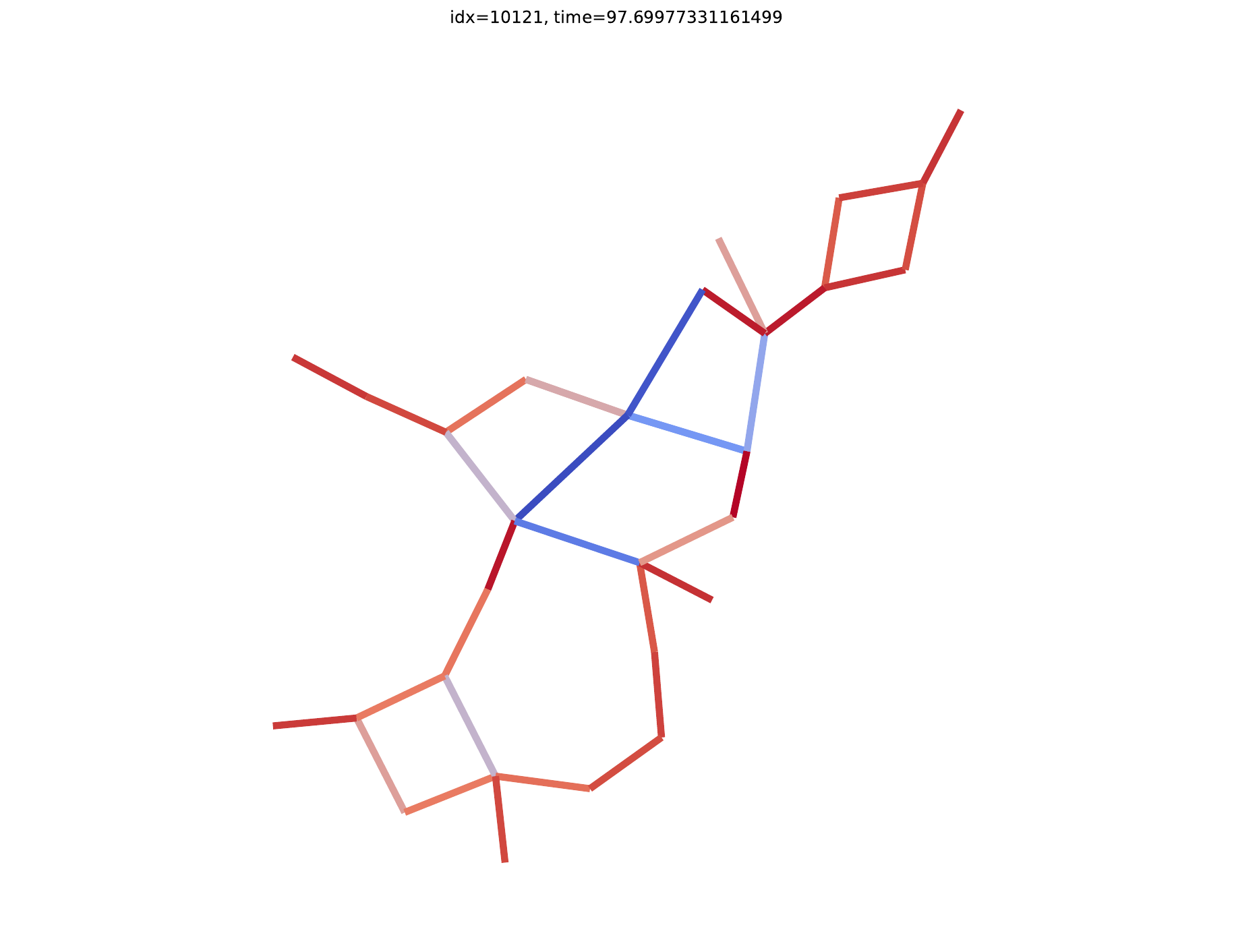} &
\imgcell{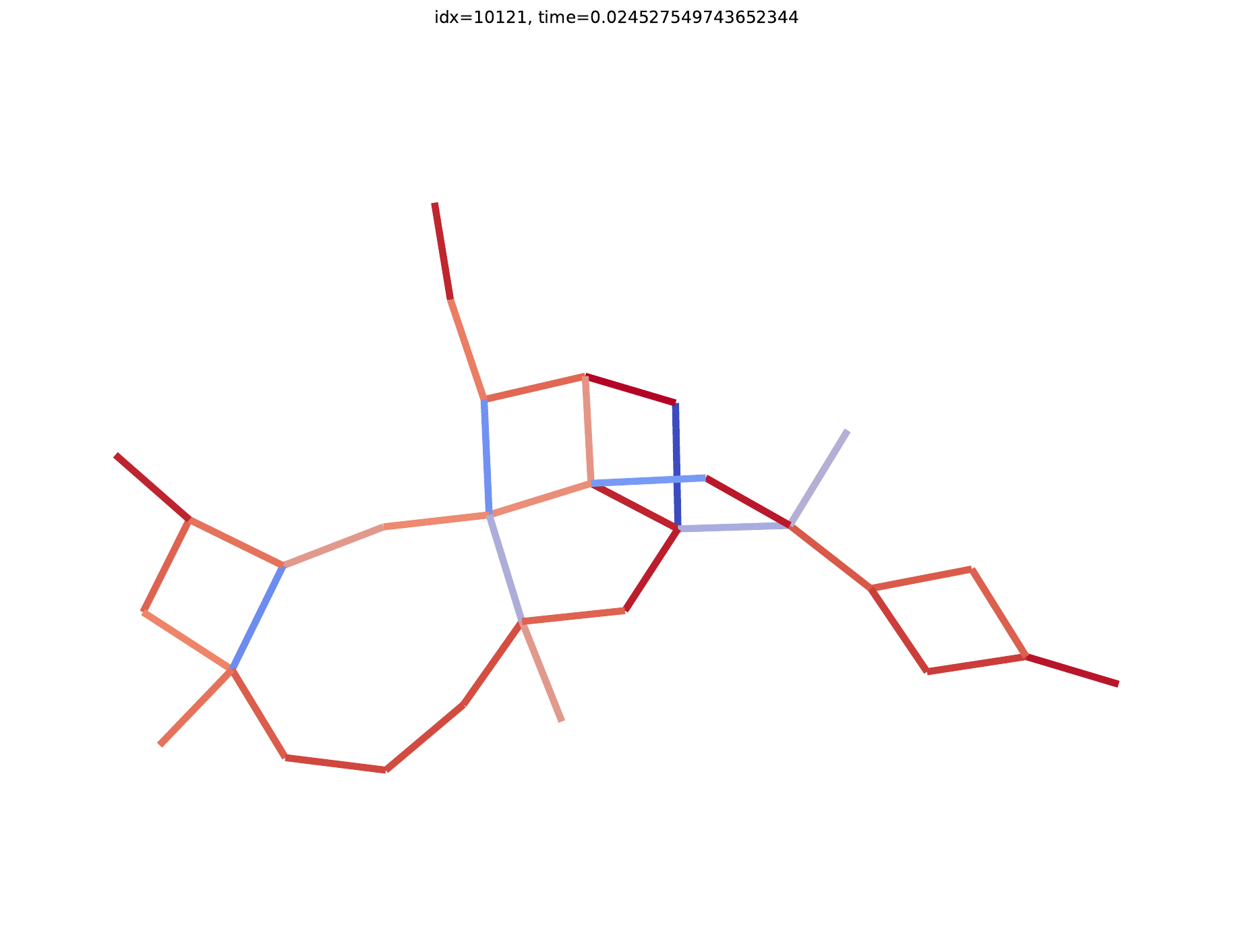} &
\imgcell{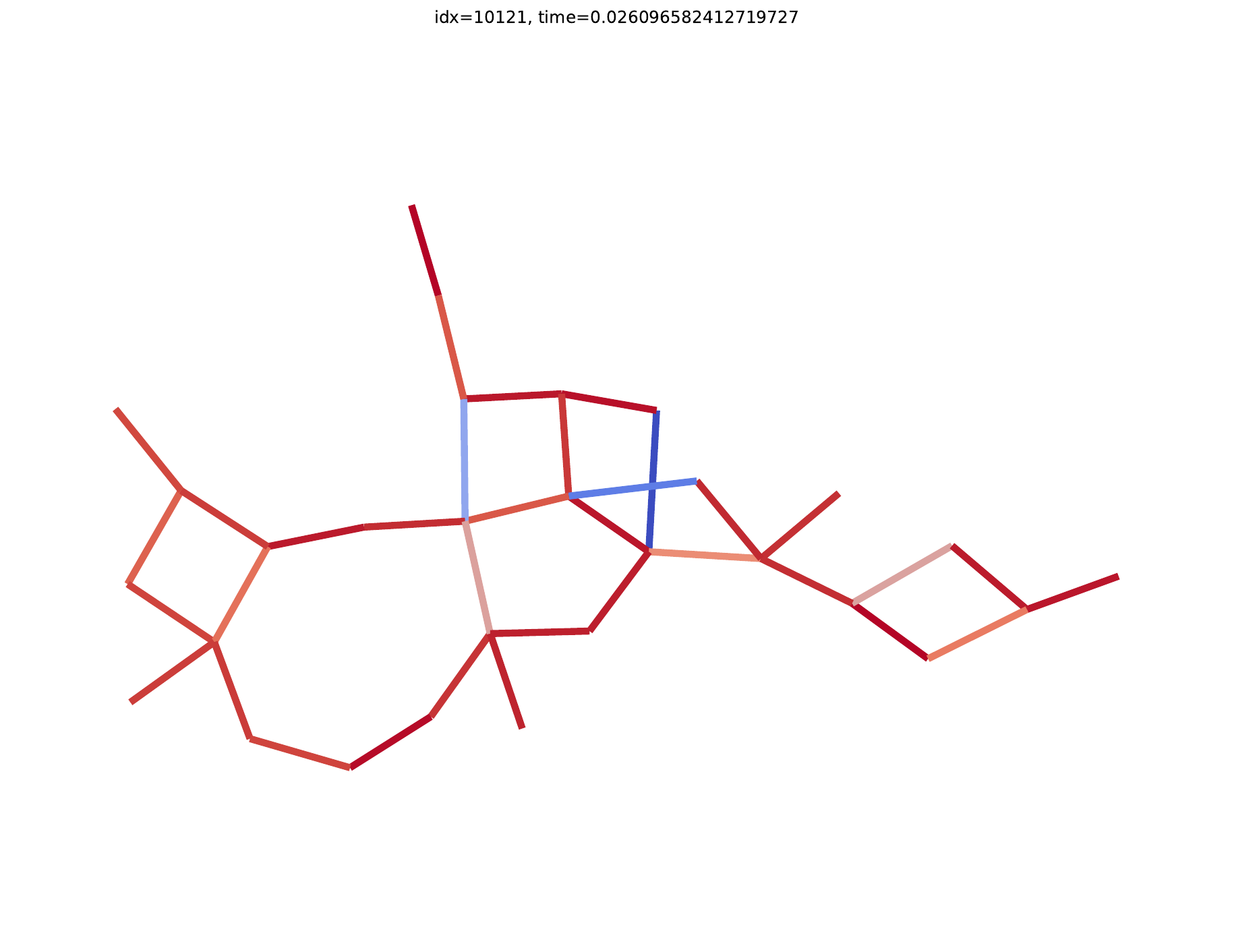} &
\imgcell{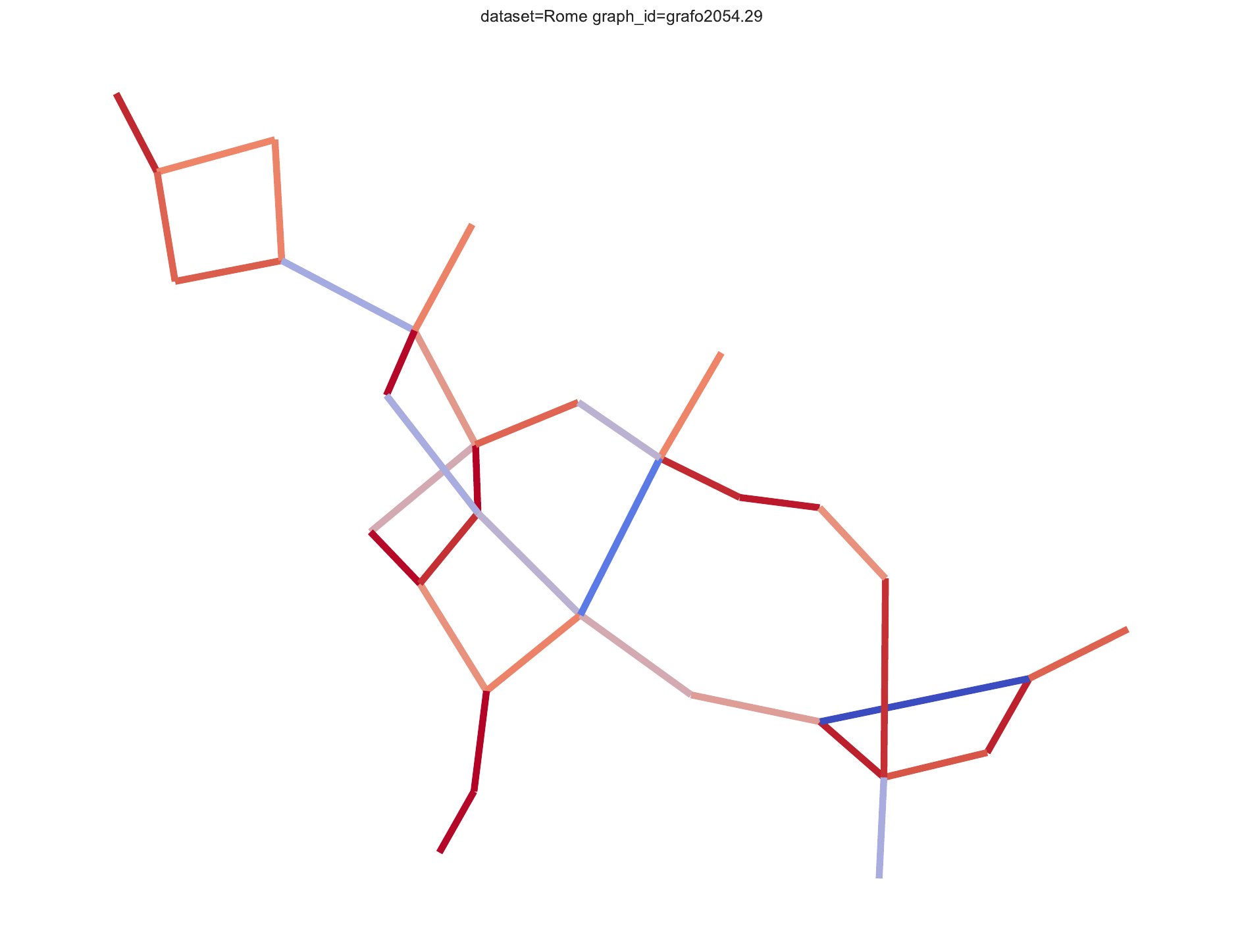} &
\imgcell{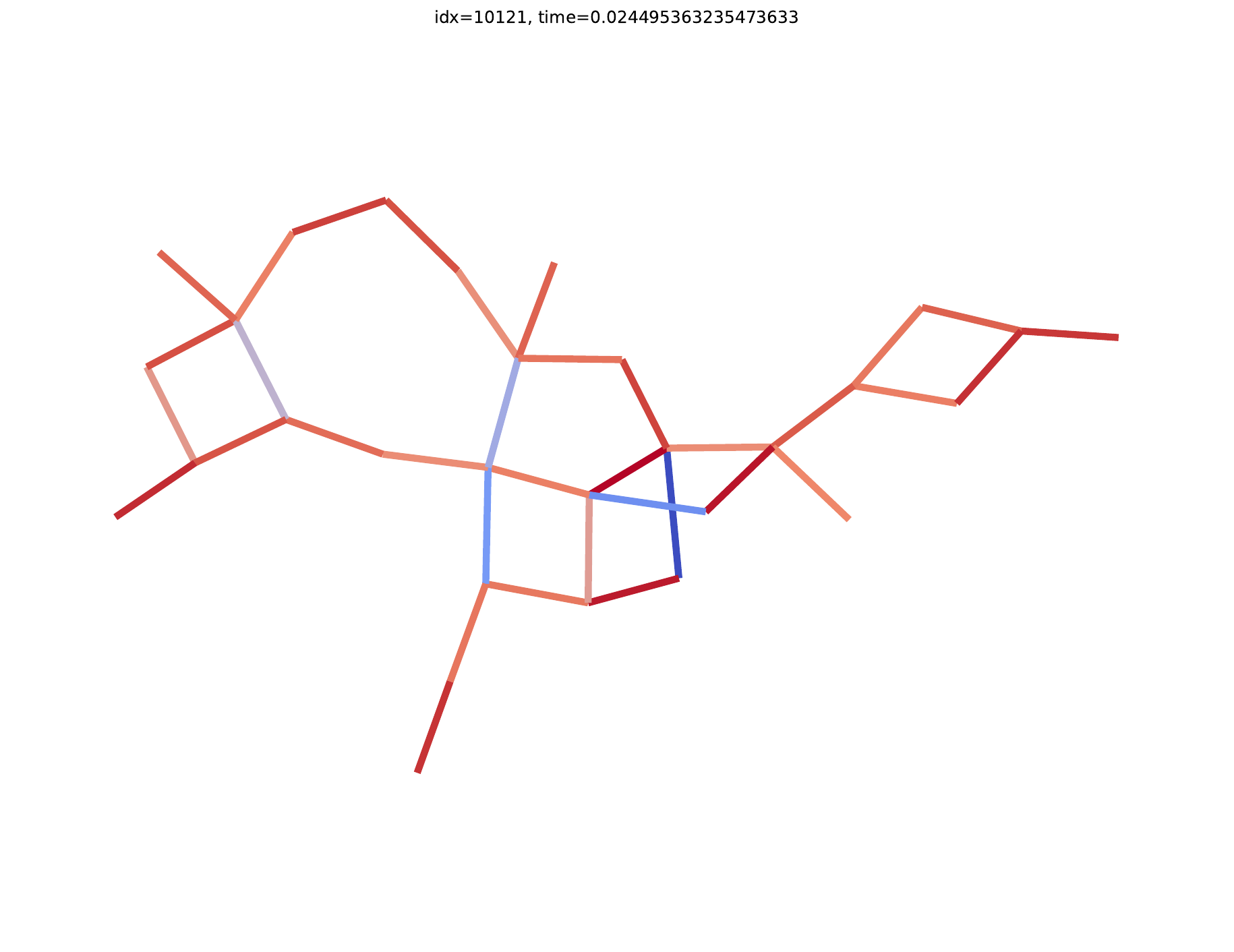} &
\imgcell{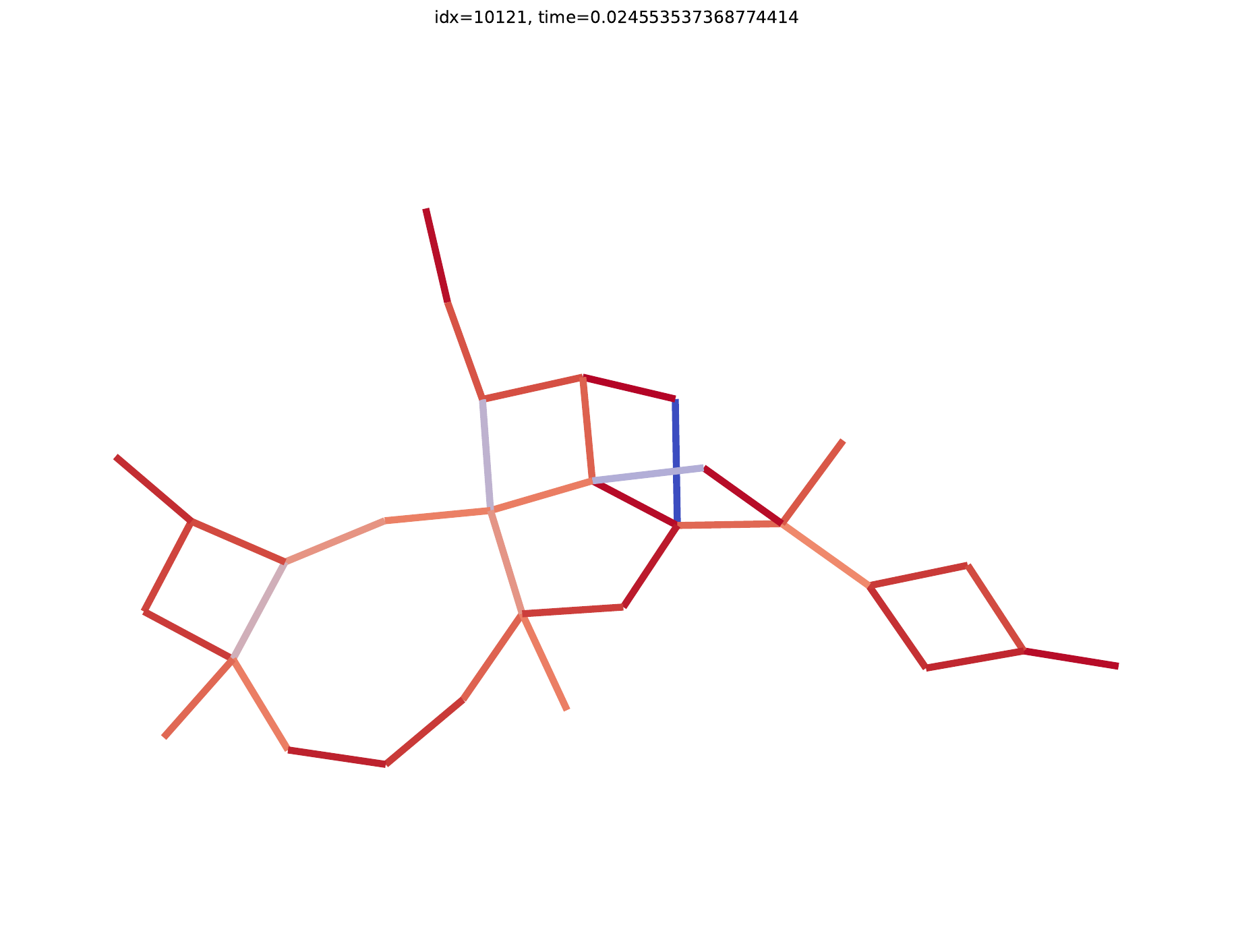} &
\imgcell{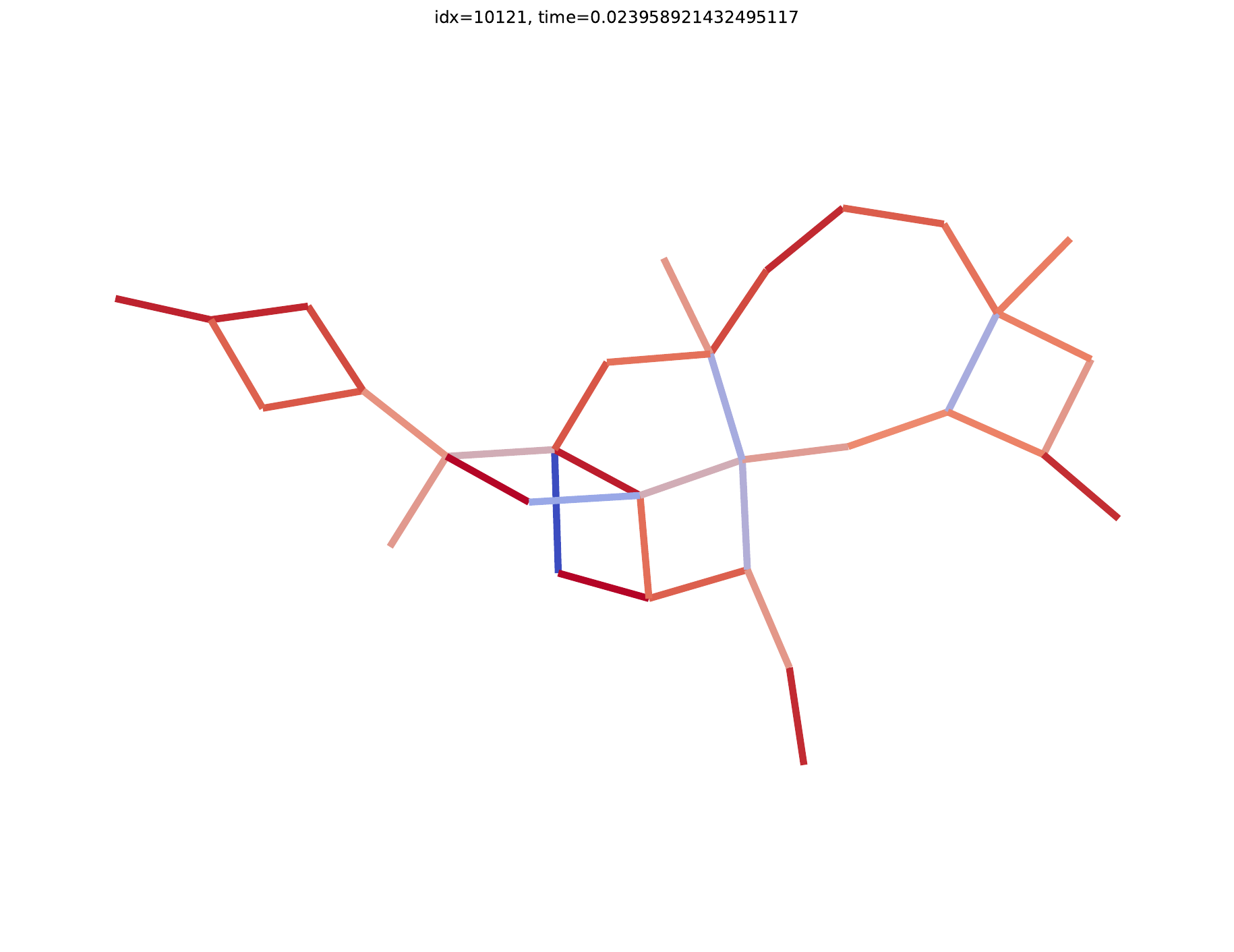} &
\imgcell{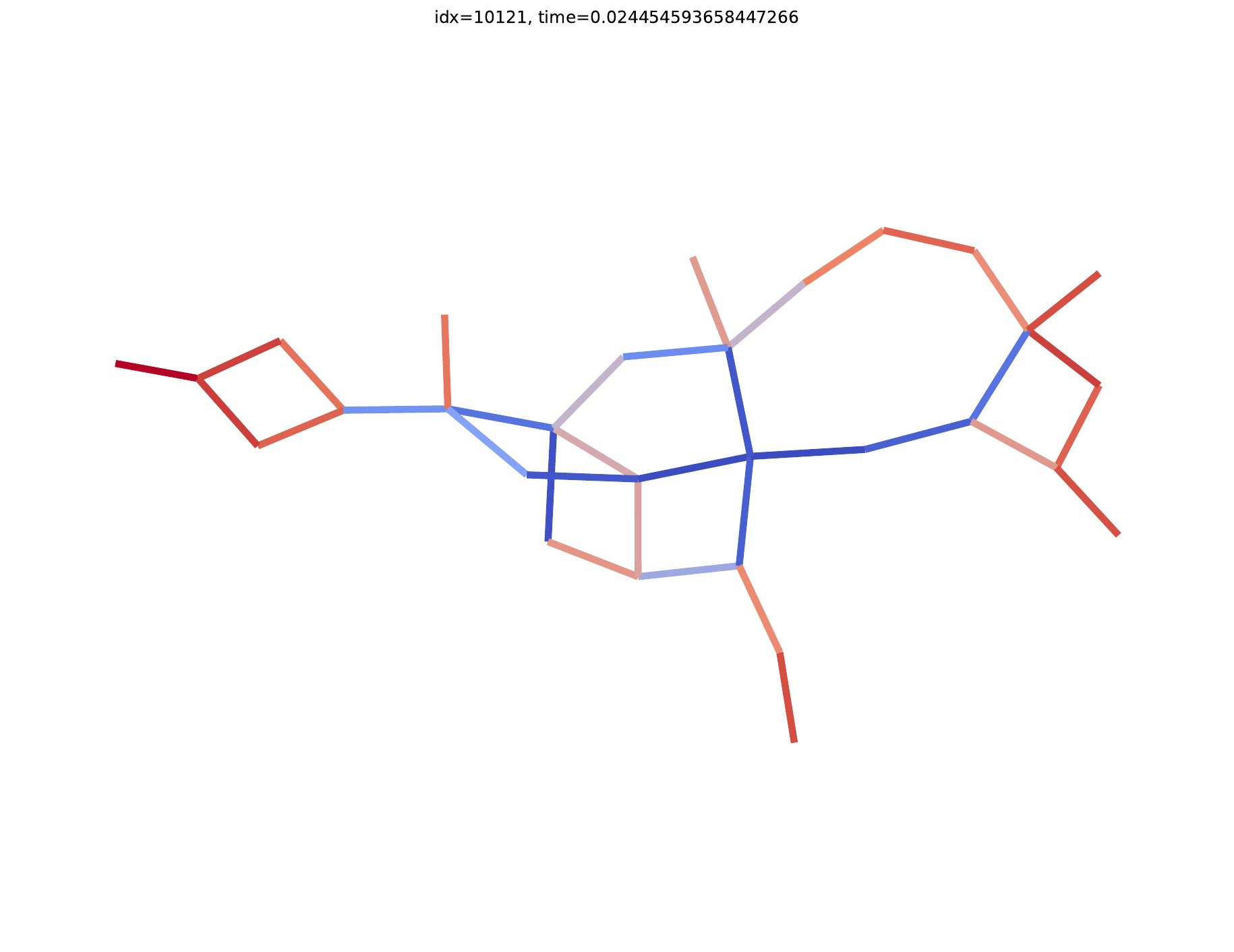} \\

&
t = 0.00s &
t = 0.31s &
t = 0.07s &
t = 0.04s &
t = 97.70s &
t = 0.02s &
t = 0.03s &
t = 0.02s &
t = 0.02s &
t = 0.02s &
t = 0.02s &
t = 0.02s \\

\makecell{\bfseries grafo9545.75\\N = 78\\M = 100} &
\imgcell{figures/rome_graphs/10130_sgd2.pdf} &
\imgcell{figures/rome_graphs/10130_pmds.pdf} &
\imgcell{figures/rome_graphs/10130_fa2.pdf} &
\imgcell{figures/rome_graphs/10130_deepgd.pdf} &
\imgcell{figures/rome_graphs/10130_gd2_stress_xing.pdf} &
\imgcell{figures/rome_graphs/10130_smartgd_stress.pdf} &
\imgcell{figures/rome_graphs/10130_smartgd_xing.pdf} &
\imgcell{figures/rome_graphs/10130_smartgd_xing_nsc.pdf} &
\imgcell{figures/rome_graphs/10130_smartgd_xangle.pdf} &
\imgcell{figures/rome_graphs/10130_smartgd_stress_xing.pdf} &
\imgcell{figures/rome_graphs/10130_smartgd_stress_xangle.pdf} &
\imgcell{figures/rome_graphs/10130_smartgd_combined.pdf} \\

&
t = 0.00s &
t = 0.57s &
t = 0.40s &
t = 0.05s &
t = 101.29s &
t = 0.06s &
t = 0.04s &
t = 0.06s &
t = 0.05s &
t = 0.04s &
t = 0.06s &
t = 0.04s \\

\makecell{\bfseries grafo3904.57\\N = 18\\M = 23} &
\imgcell{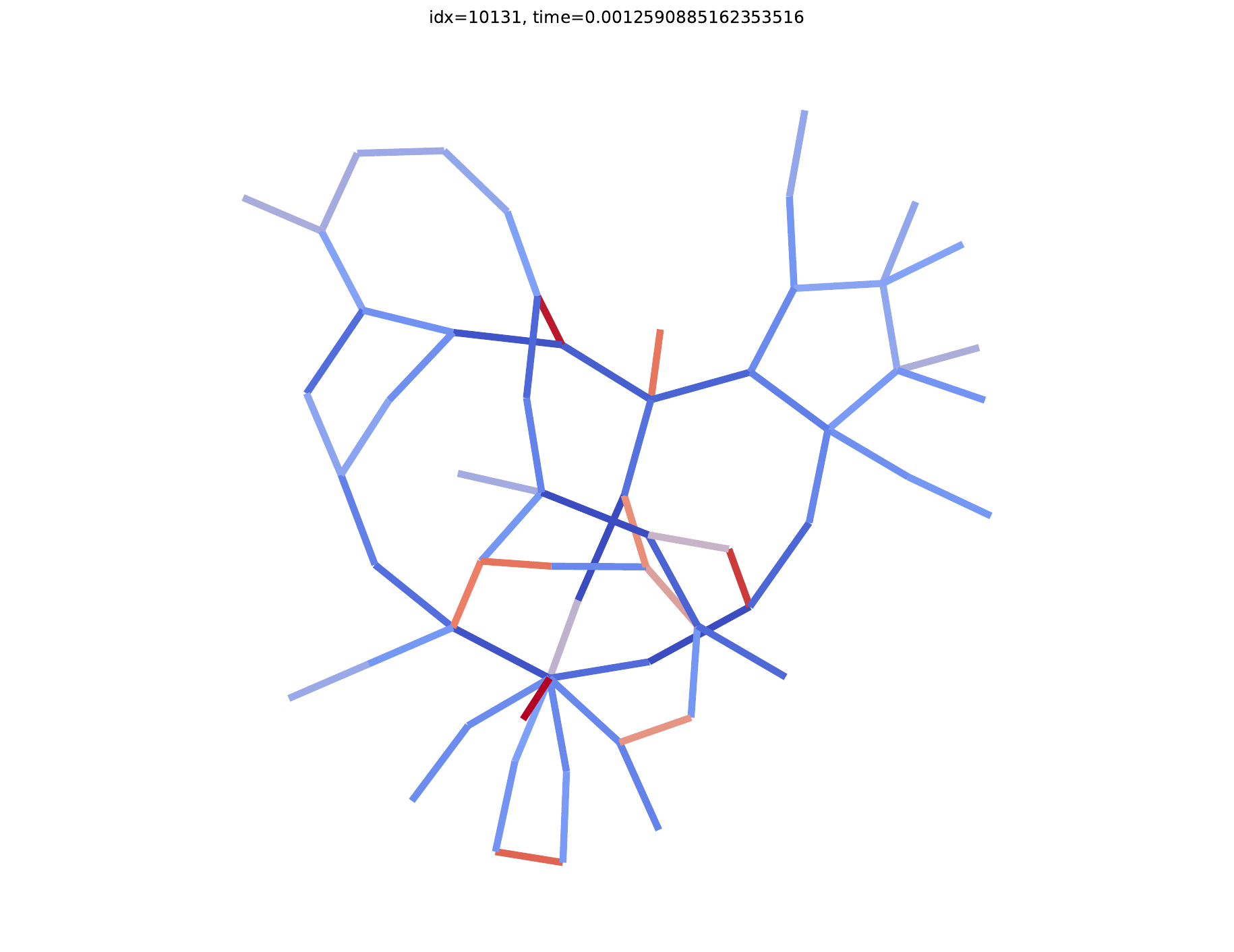} &
\imgcell{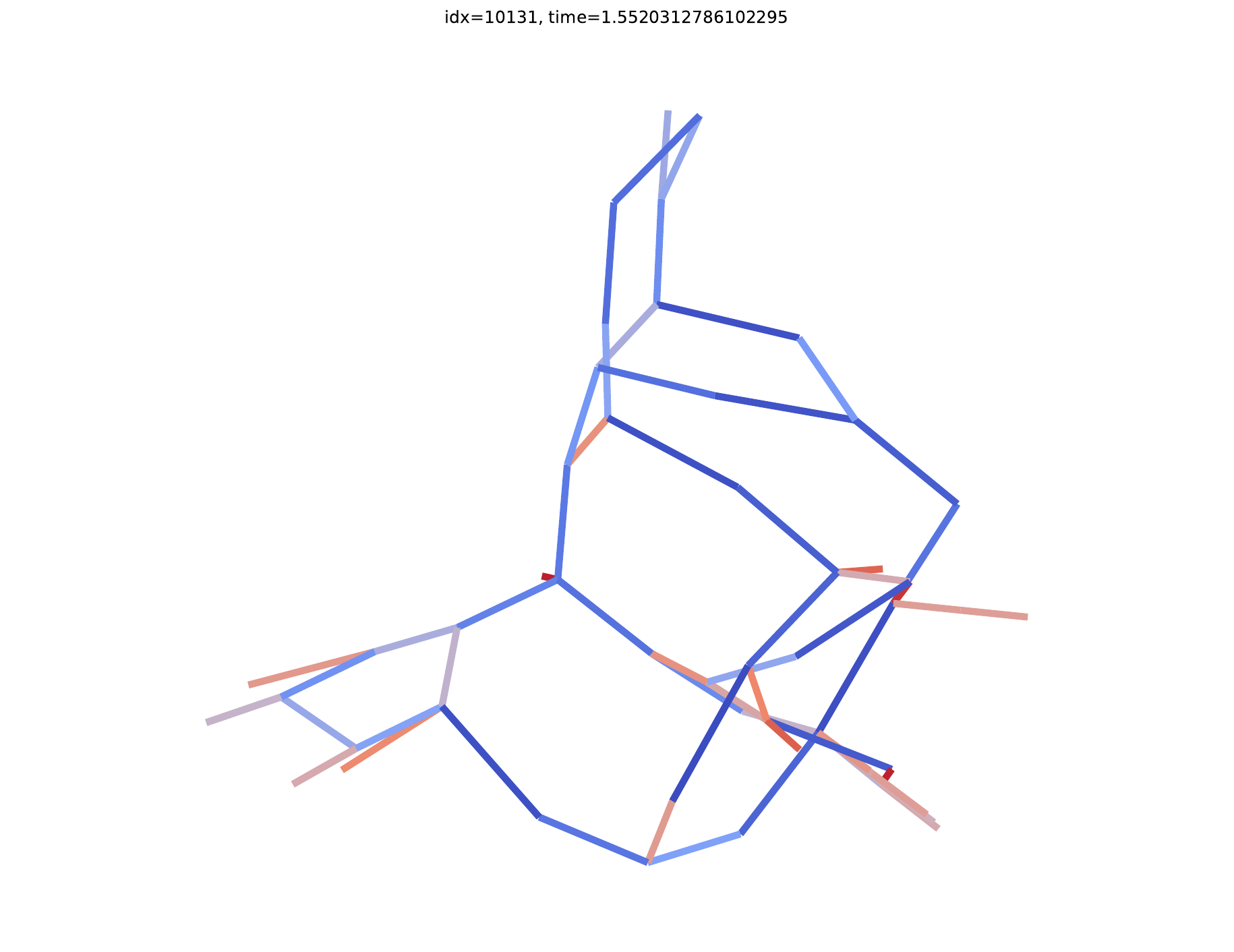} &
\imgcell{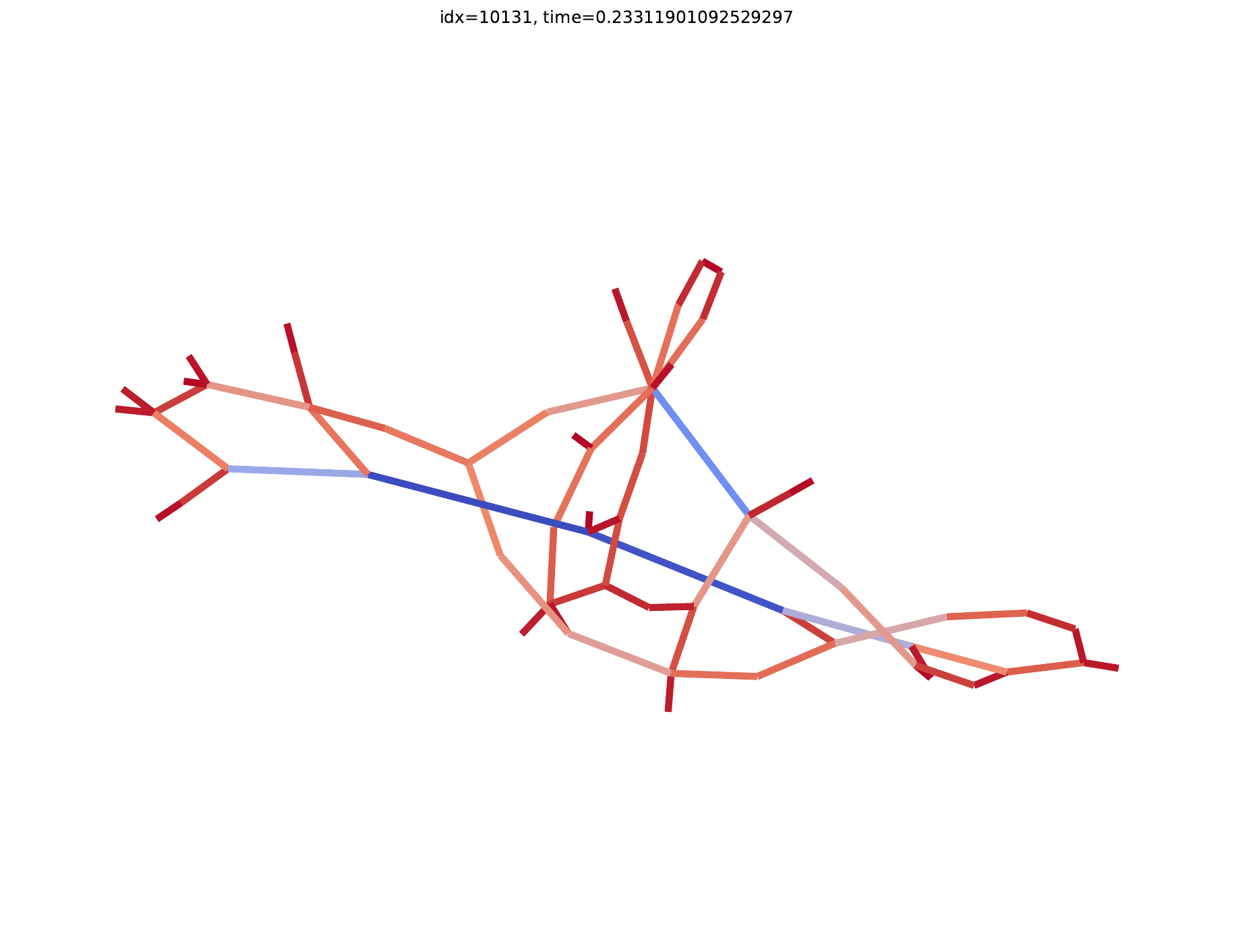} &
\imgcell{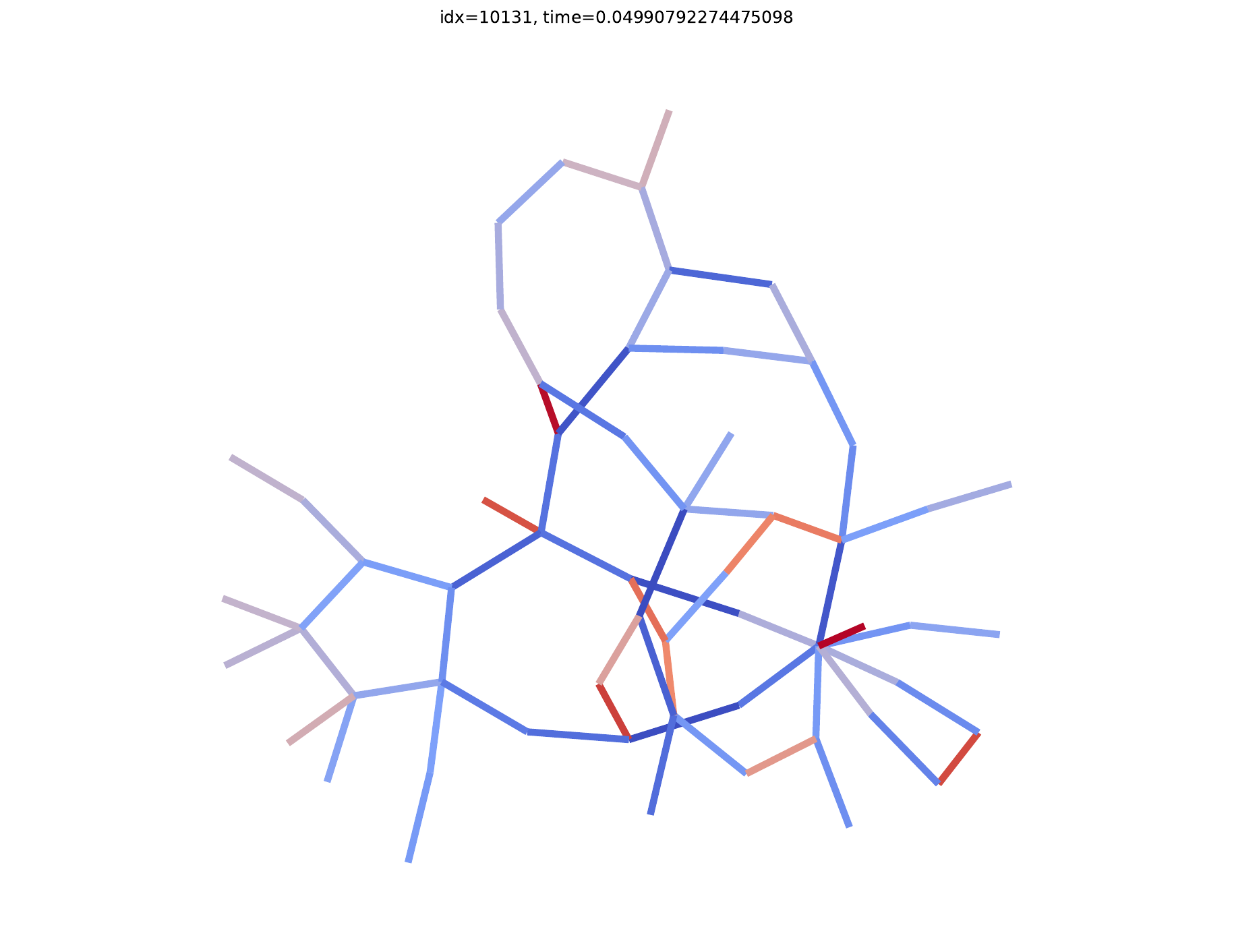} &
\imgcell{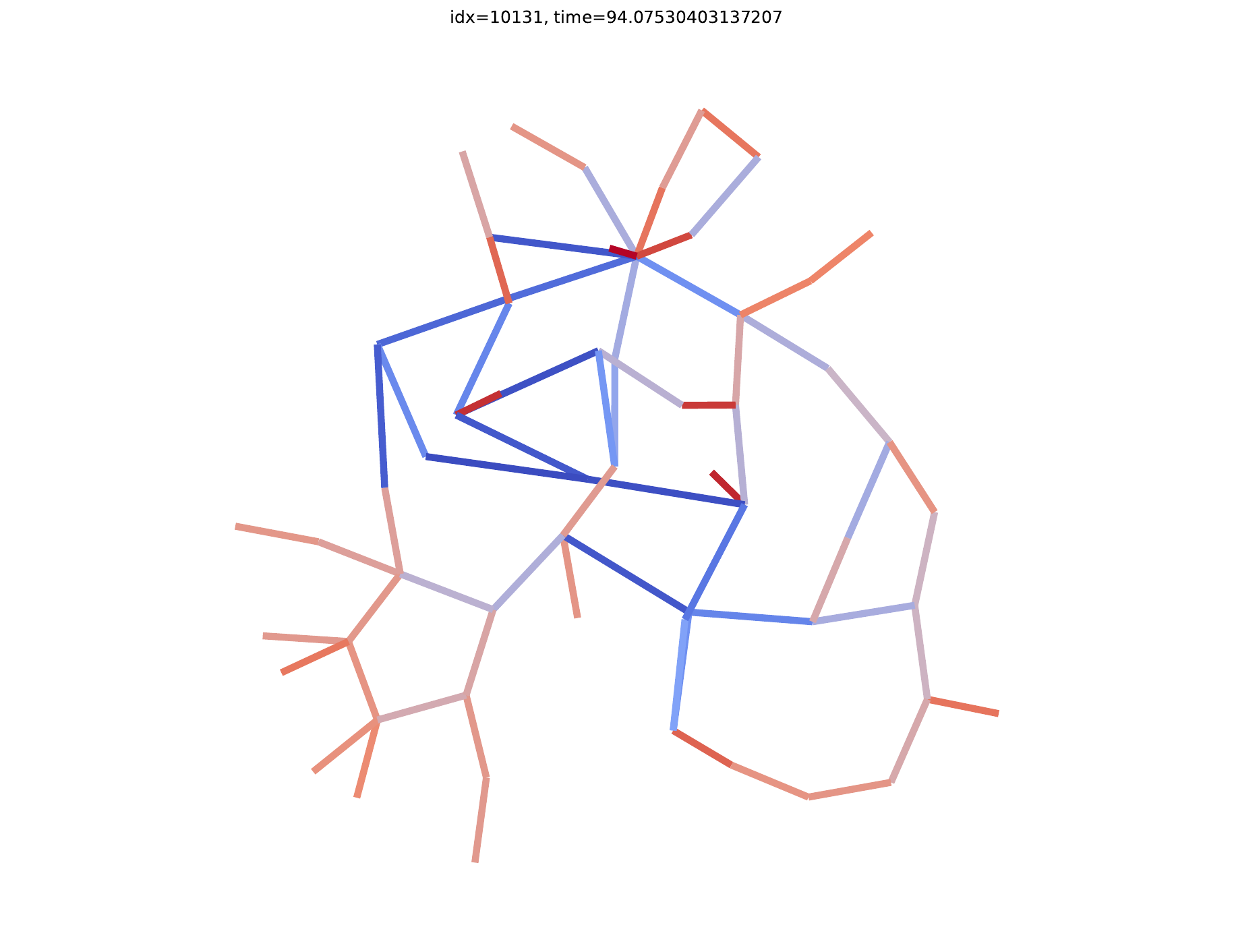} &
\imgcell{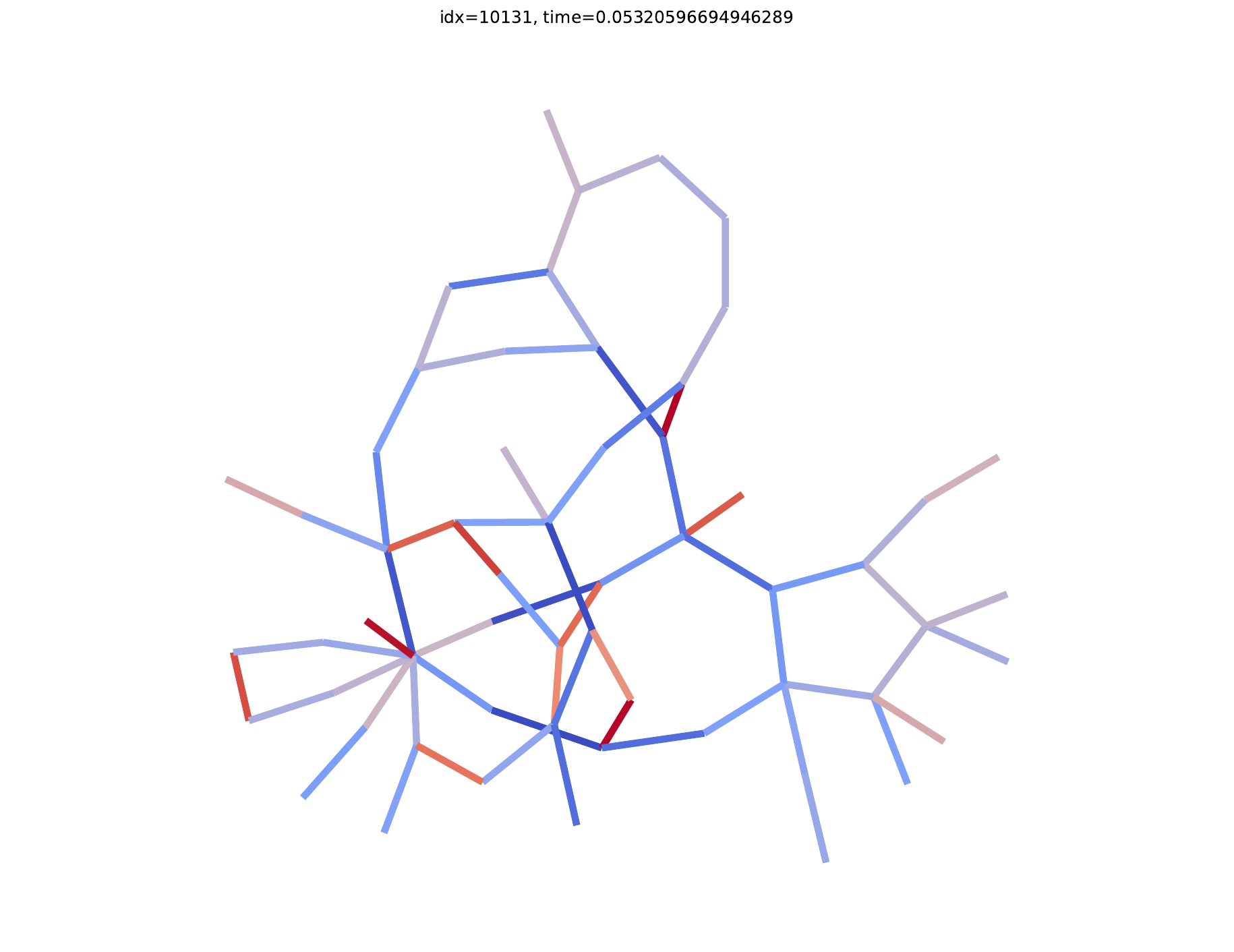} &
\imgcell{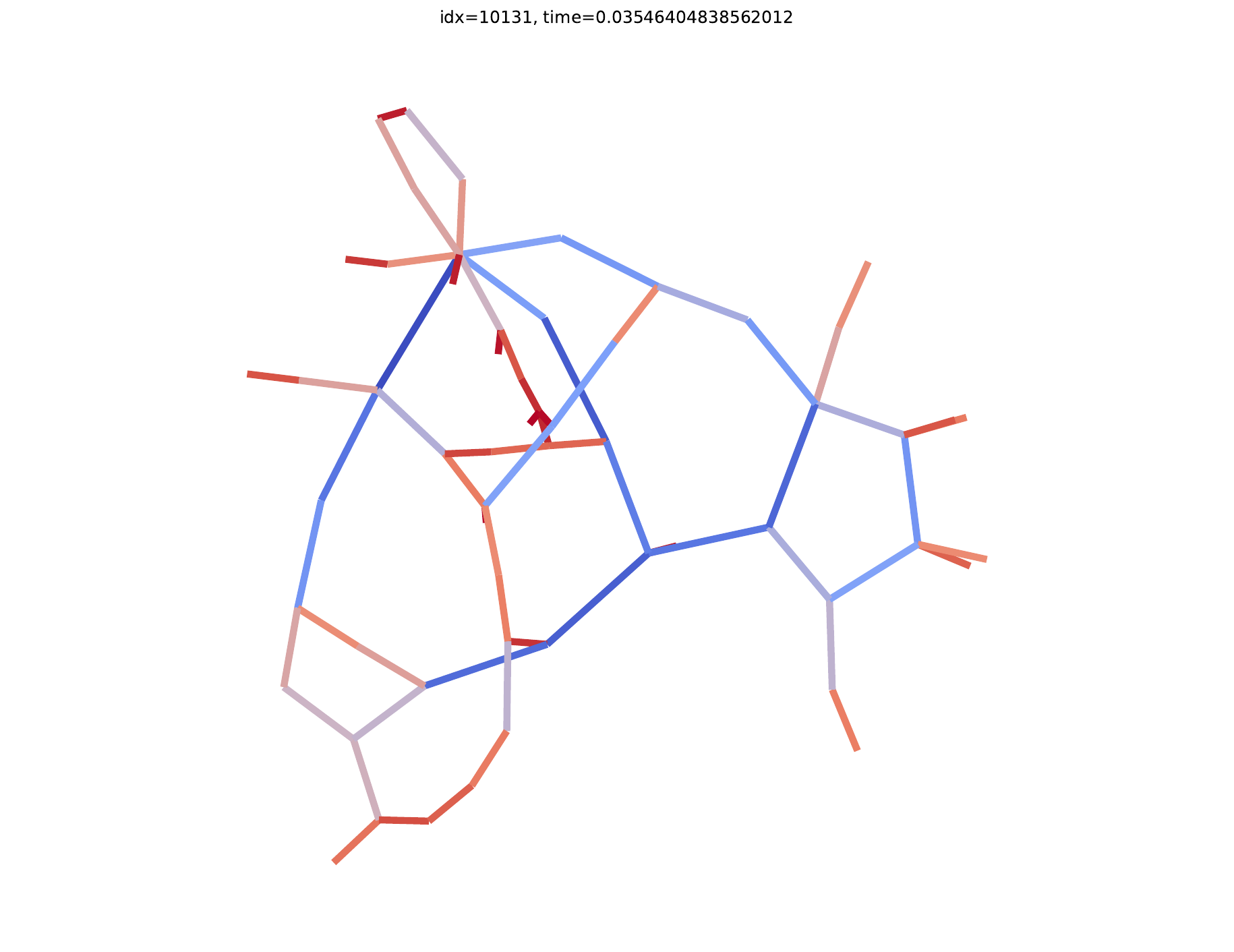} &
\imgcell{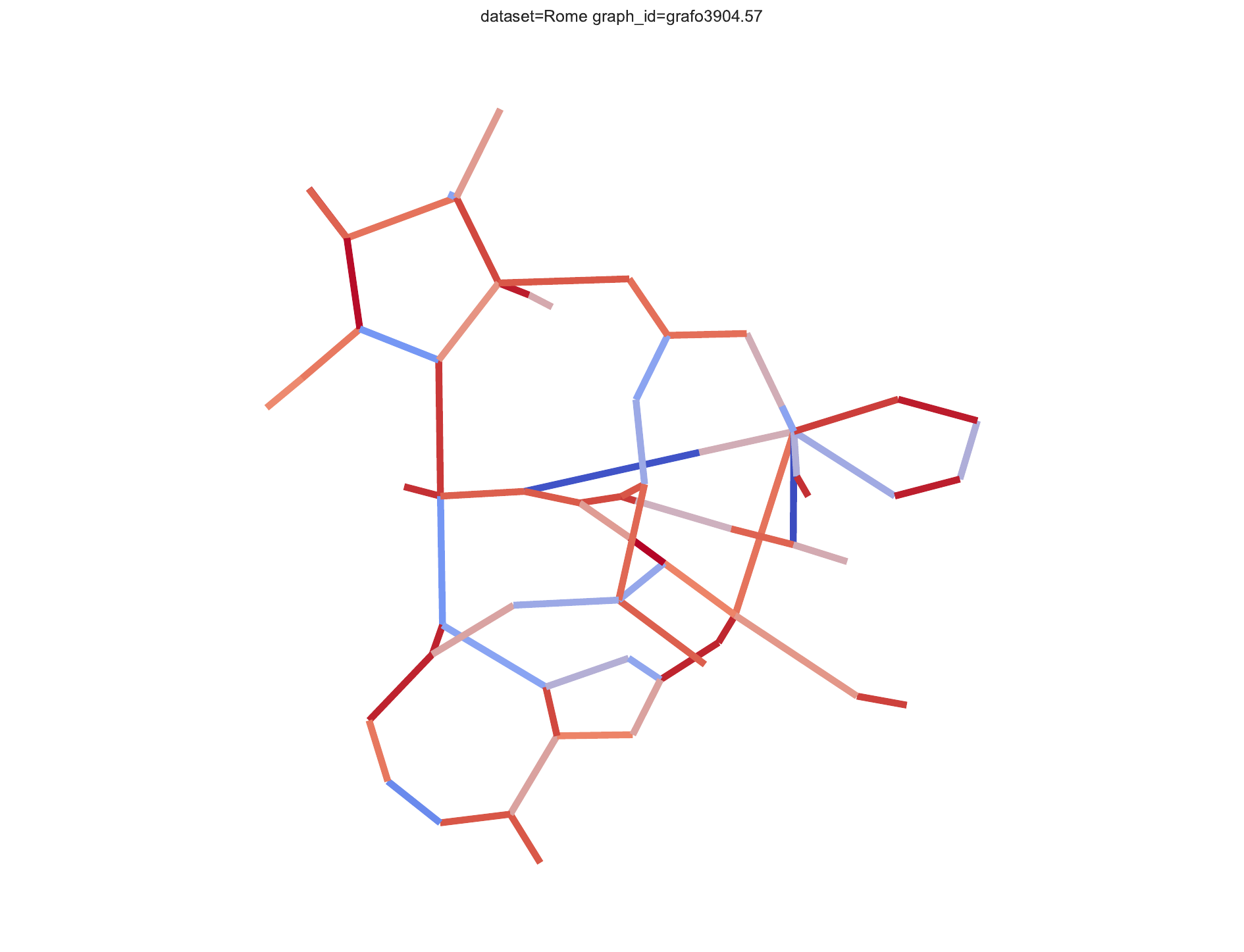} &
\imgcell{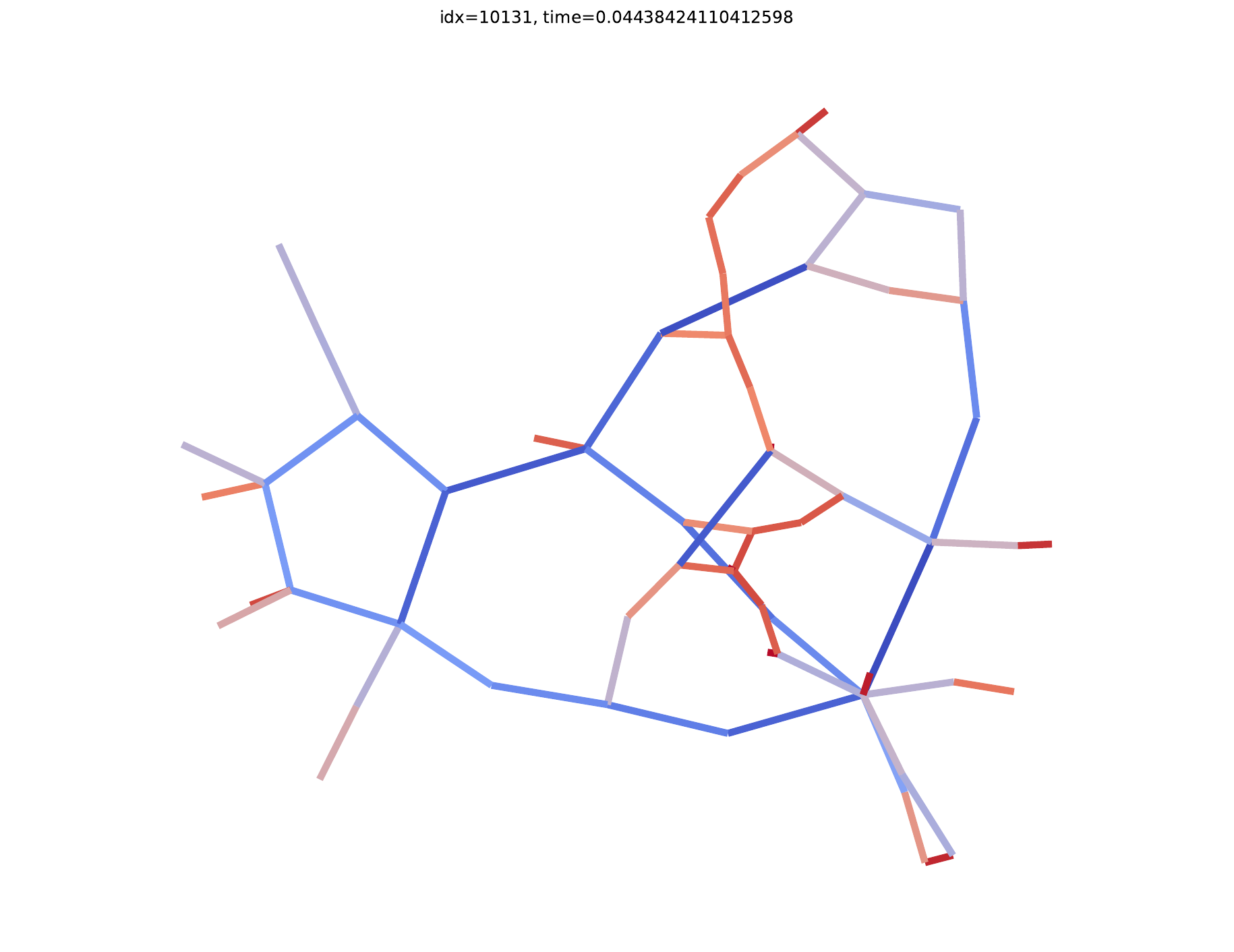} &
\imgcell{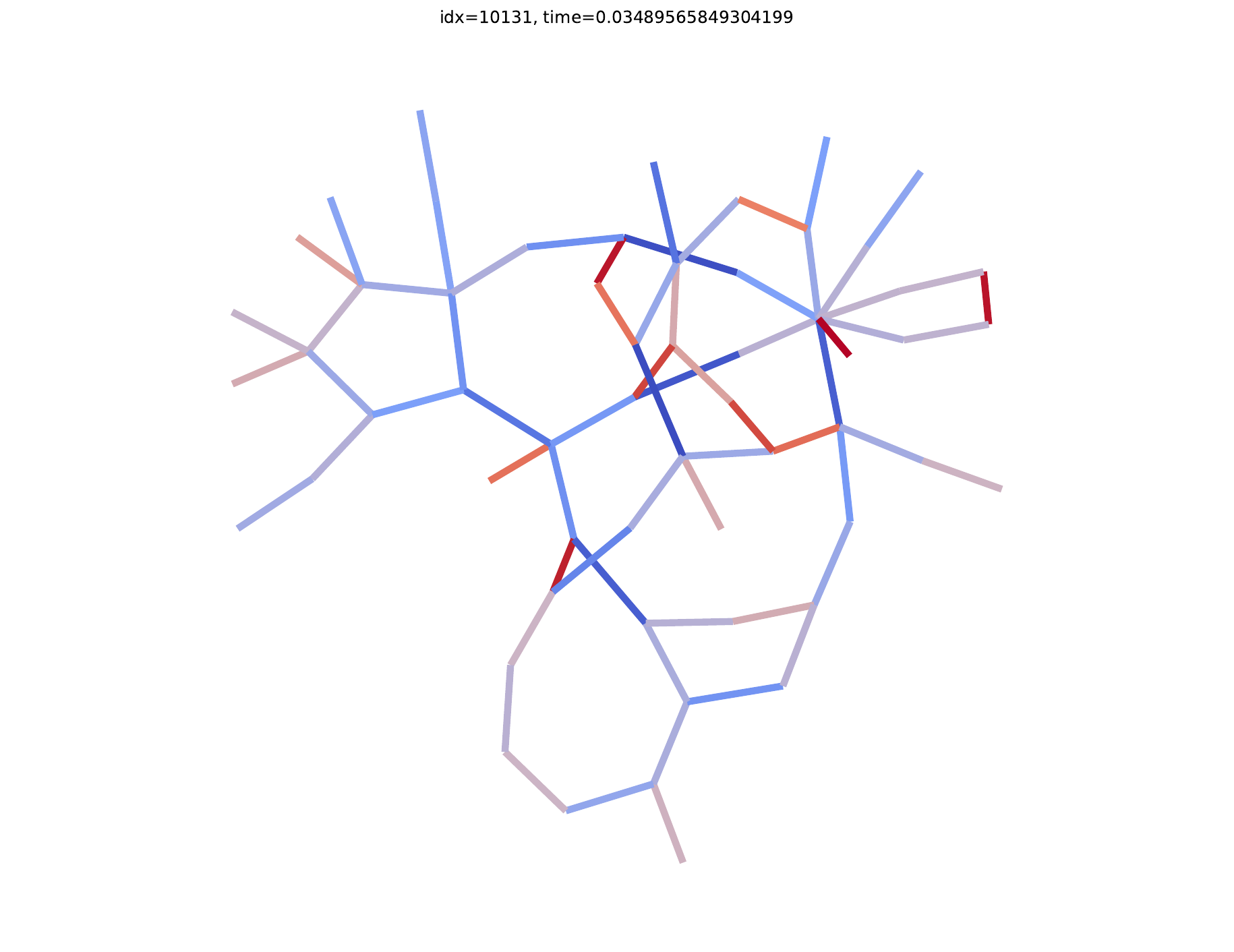} &
\imgcell{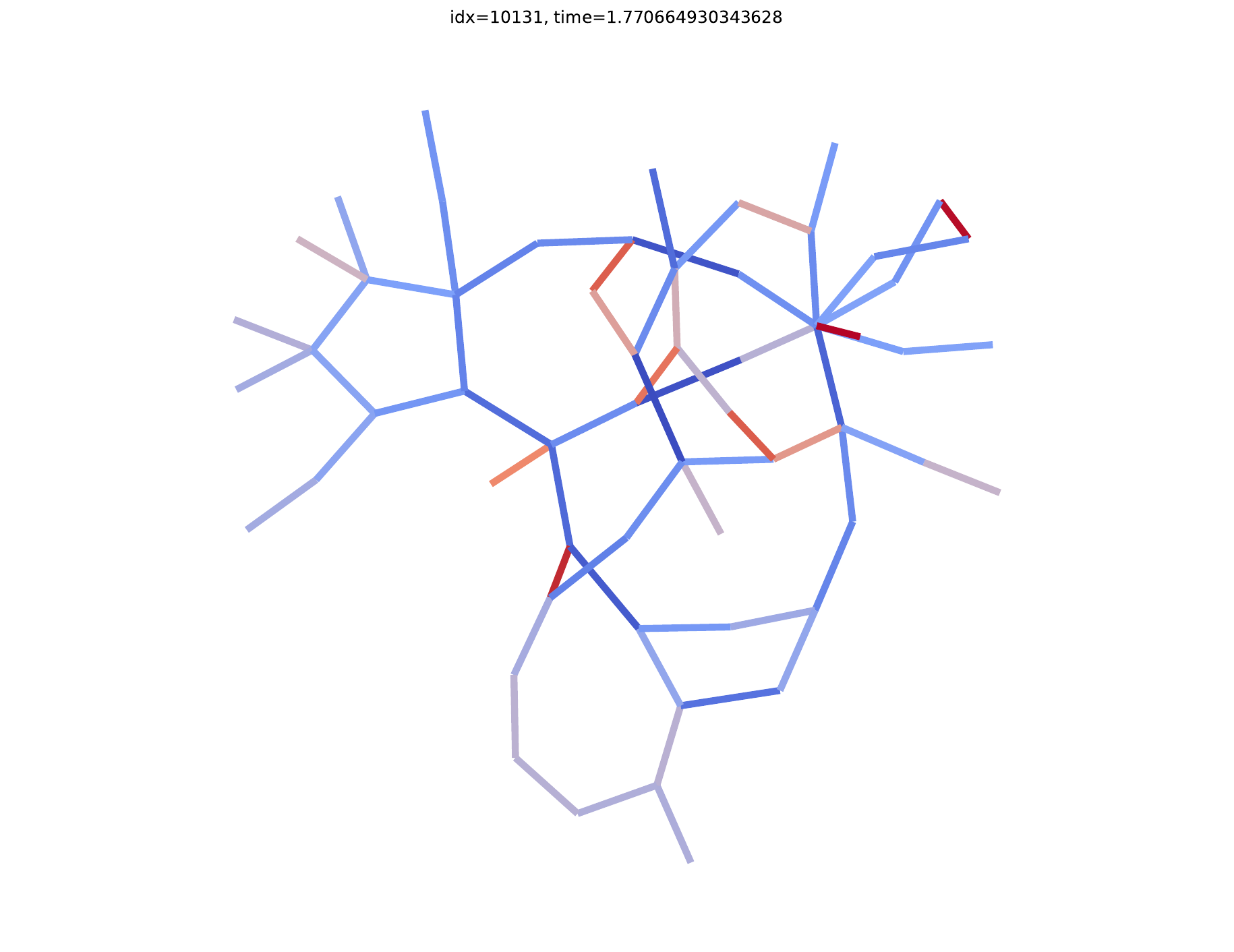} &
\imgcell{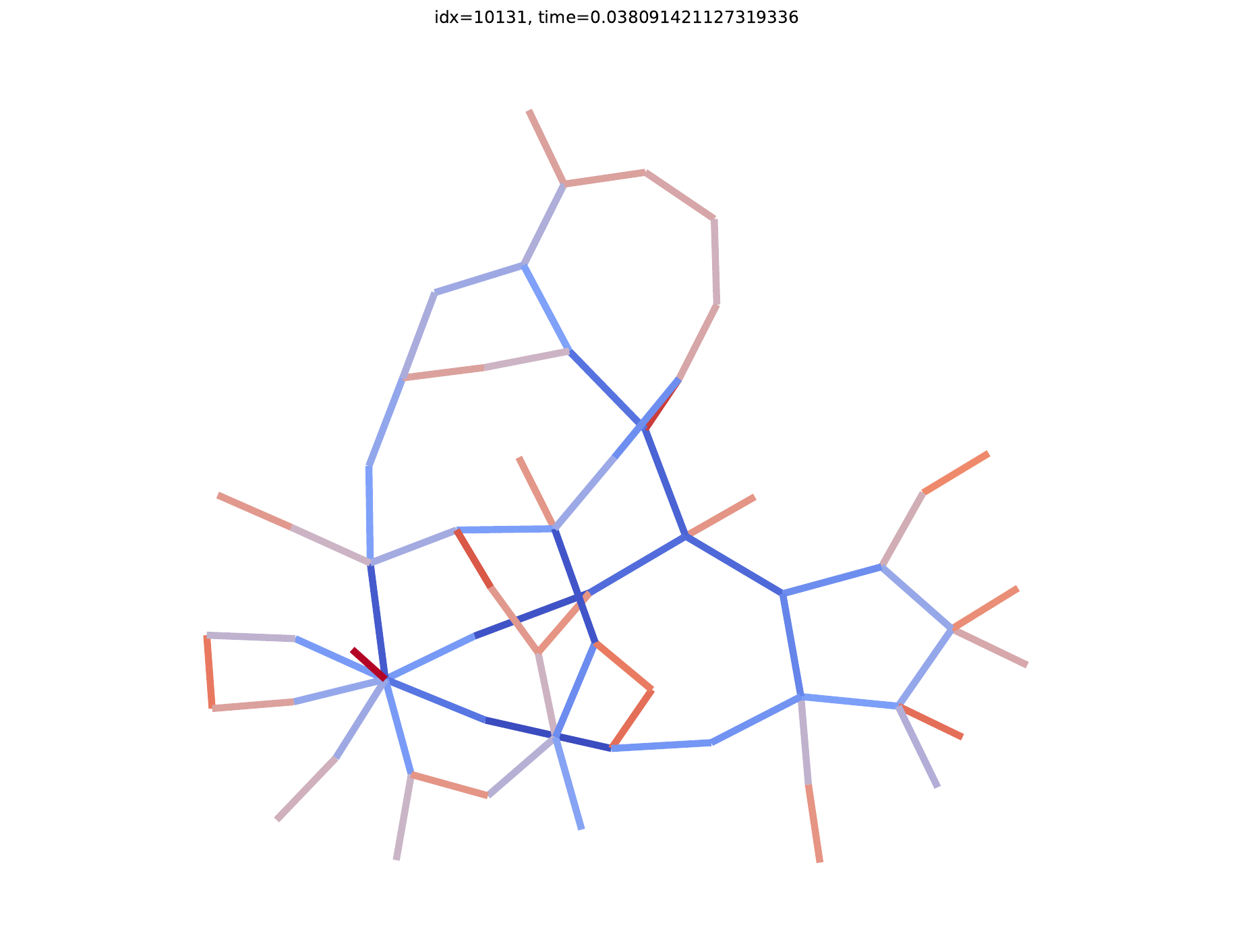} \\

&
t = 0.00s &
t = 1.55s &
t = 0.23s &
t = 0.05s &
t = 94.08s &
t = 0.05s &
t = 0.04s &
t = 0.04s &
t = 0.04s &
t = 0.03s &
t = 0.05s &
t = 0.04s \\

\makecell{\bfseries grafo3004.38\\N = 92\\M = 133} &
\imgcell{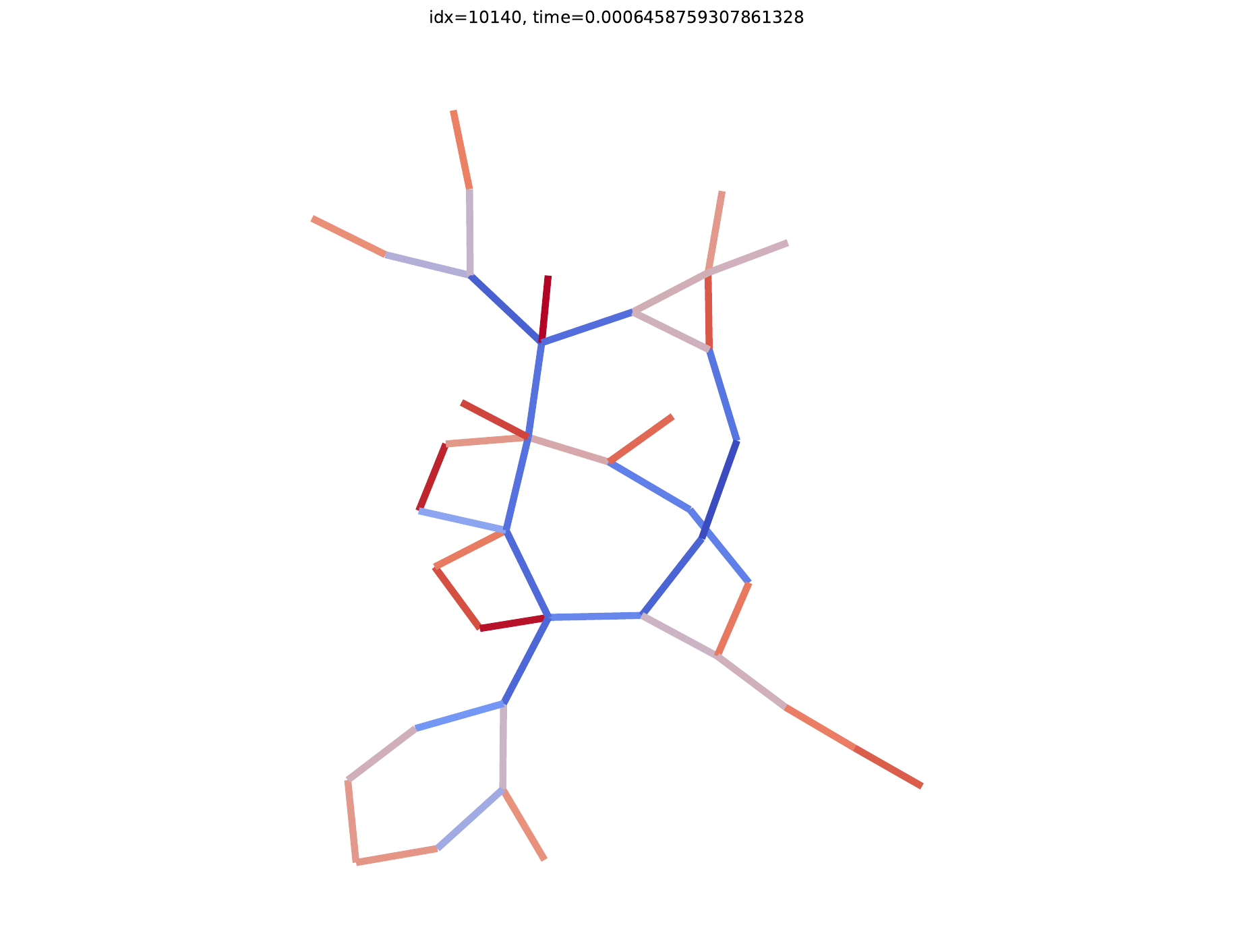} &
\imgcell{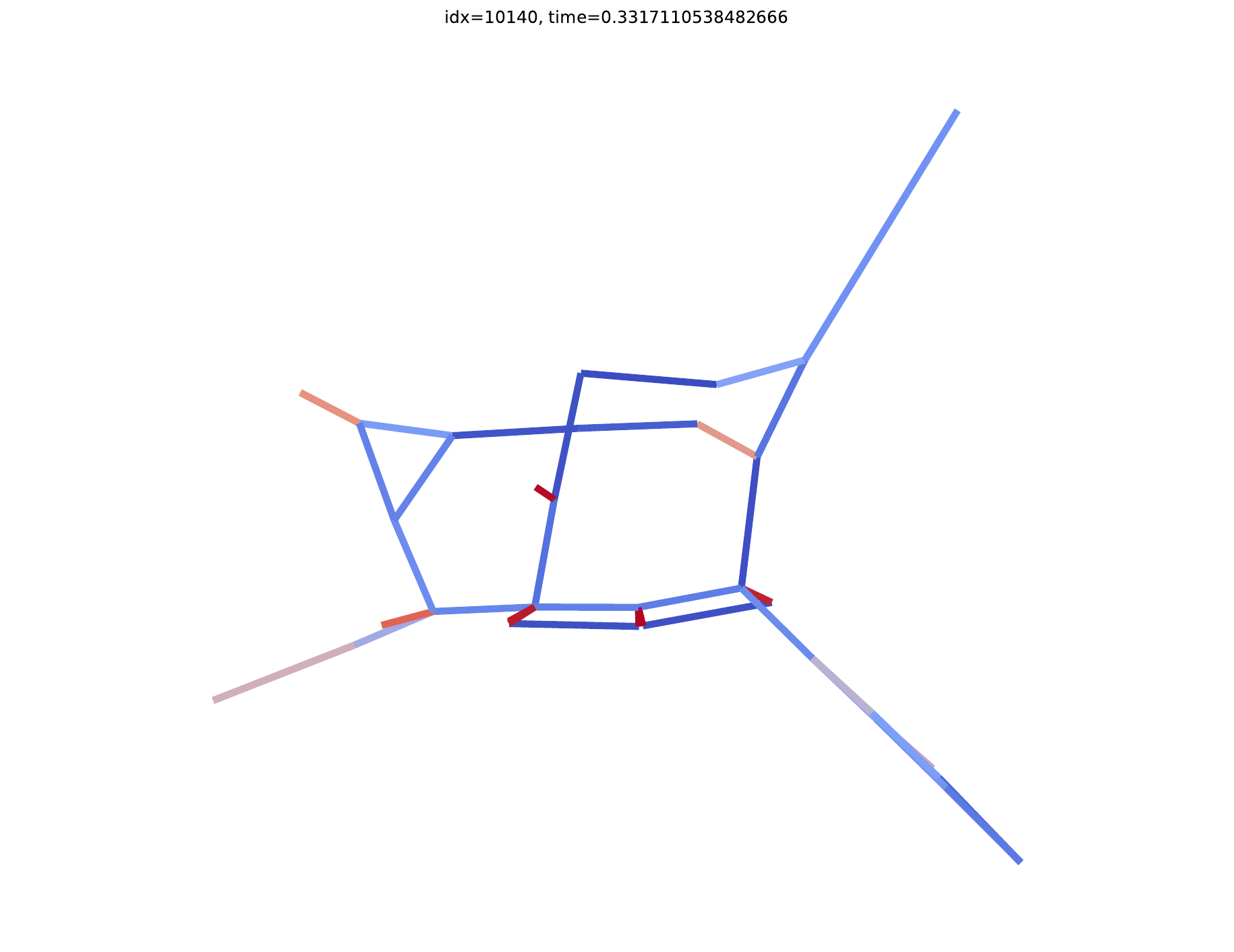} &
\imgcell{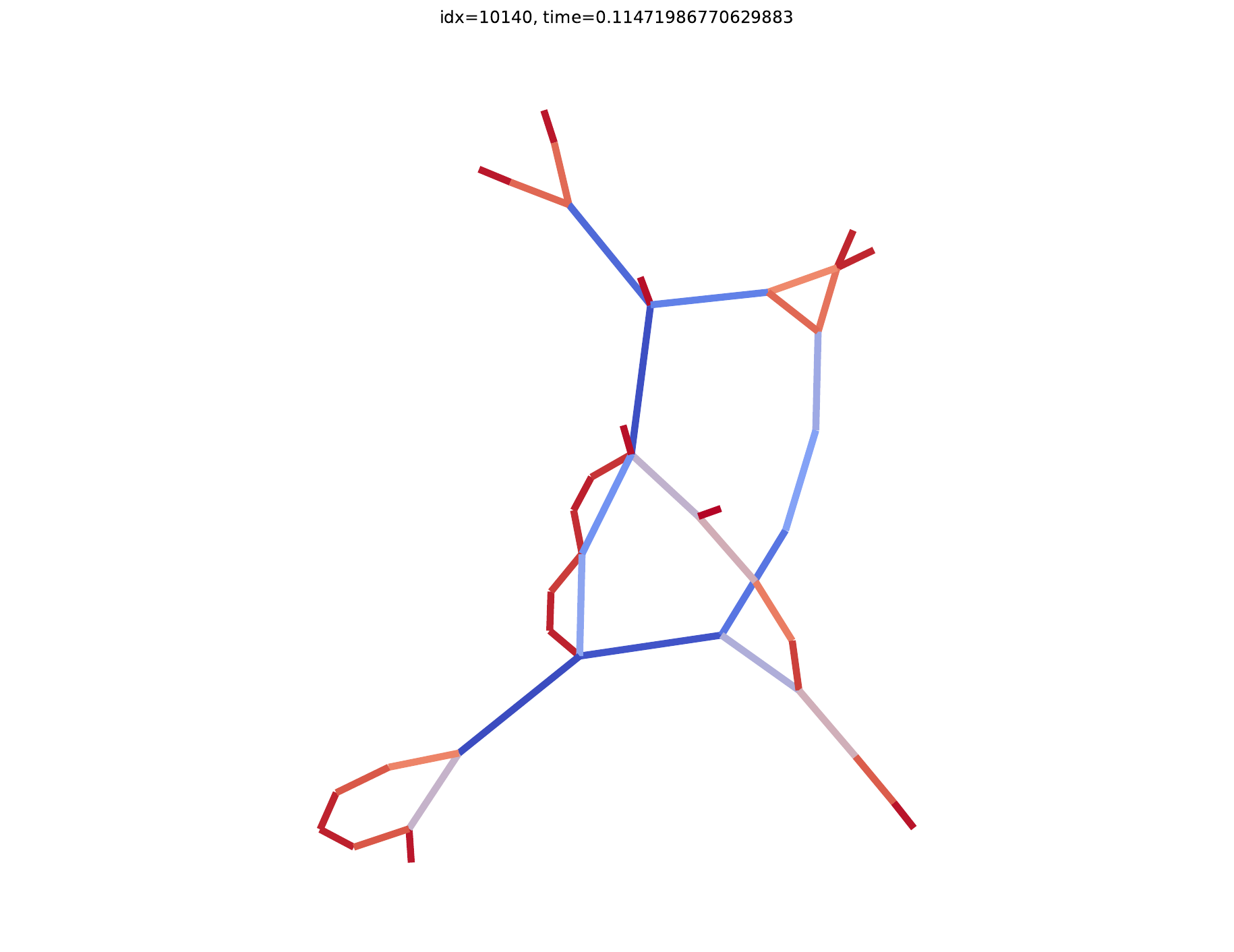} &
\imgcell{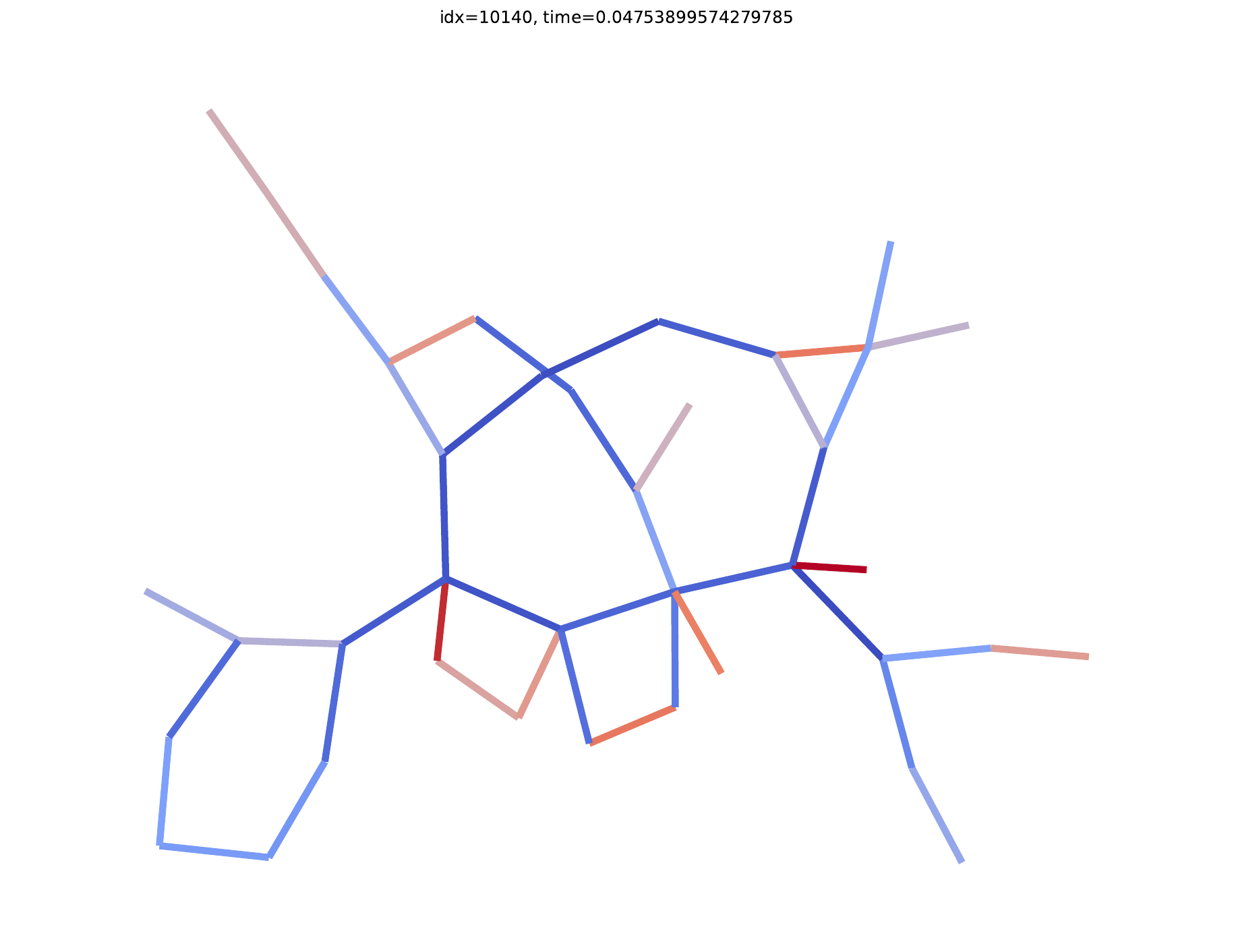} &
\imgcell{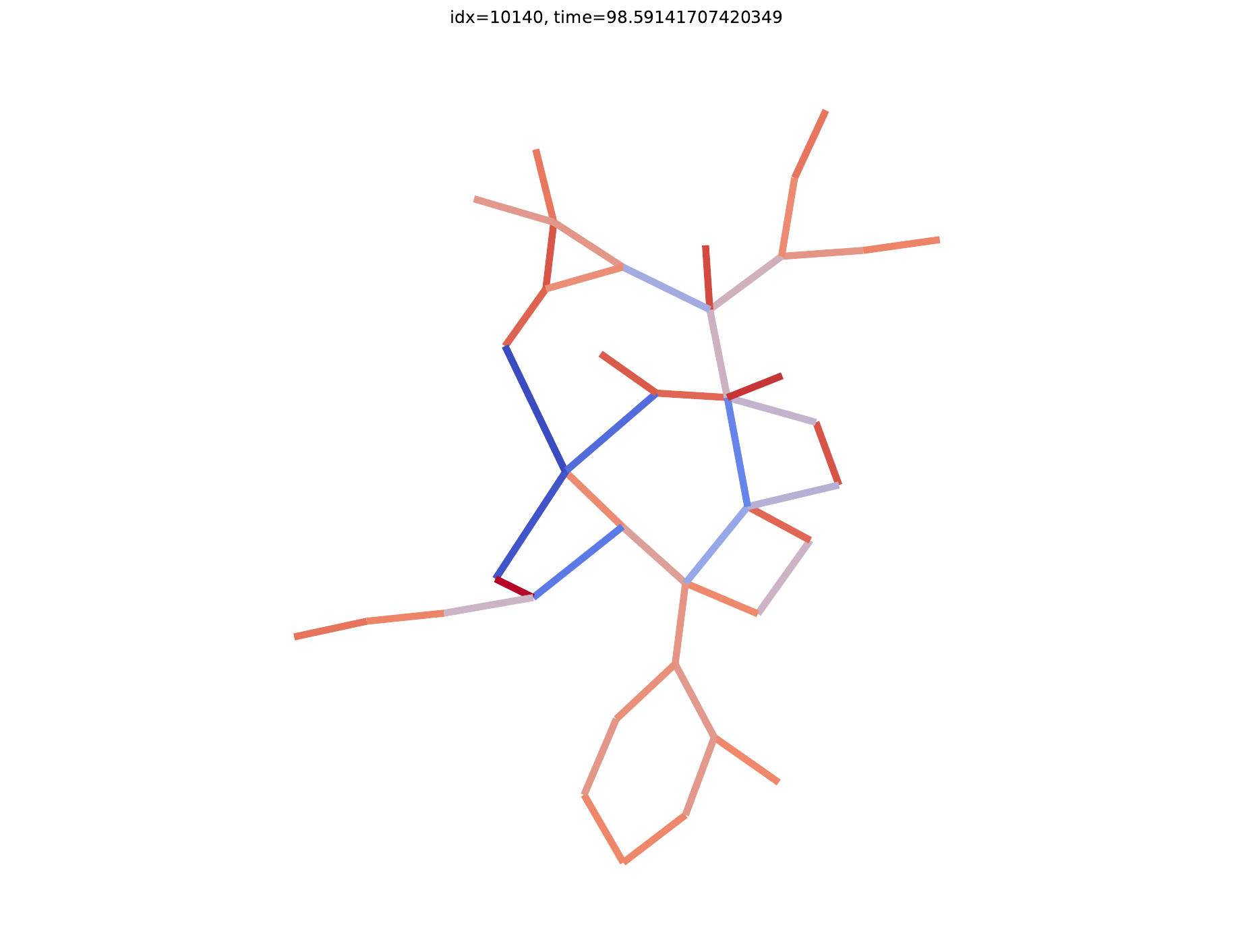} &
\imgcell{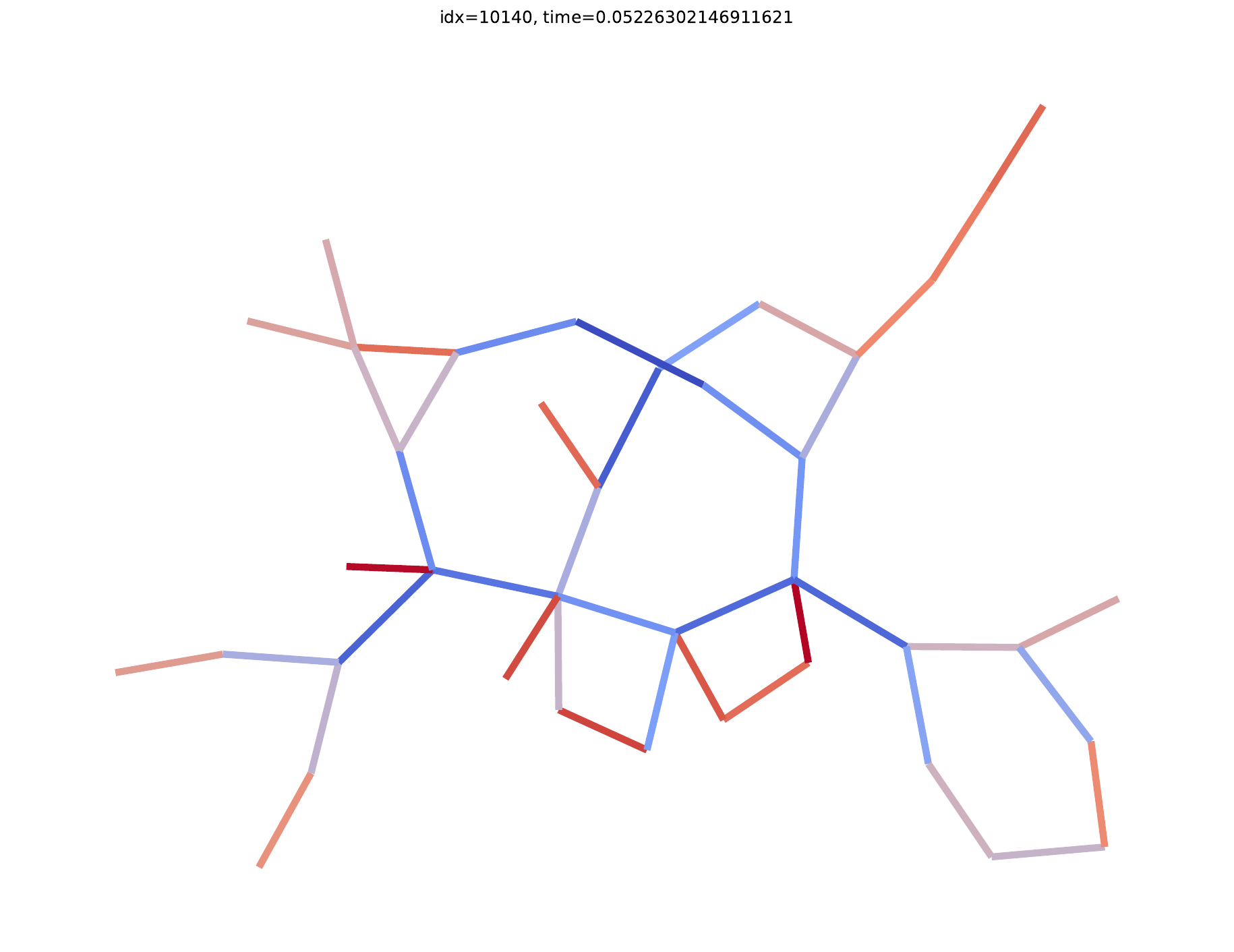} &
\imgcell{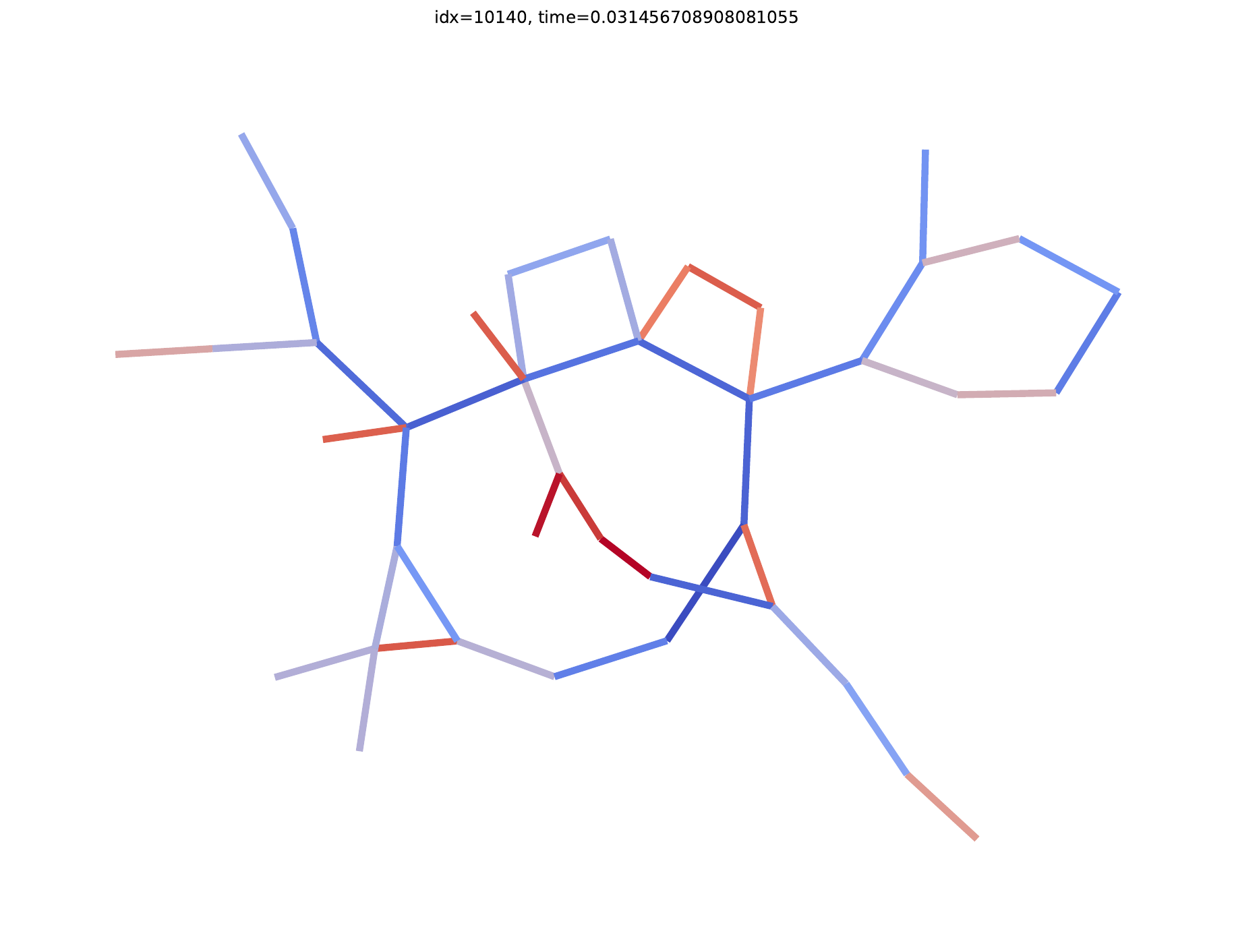} &
\imgcell{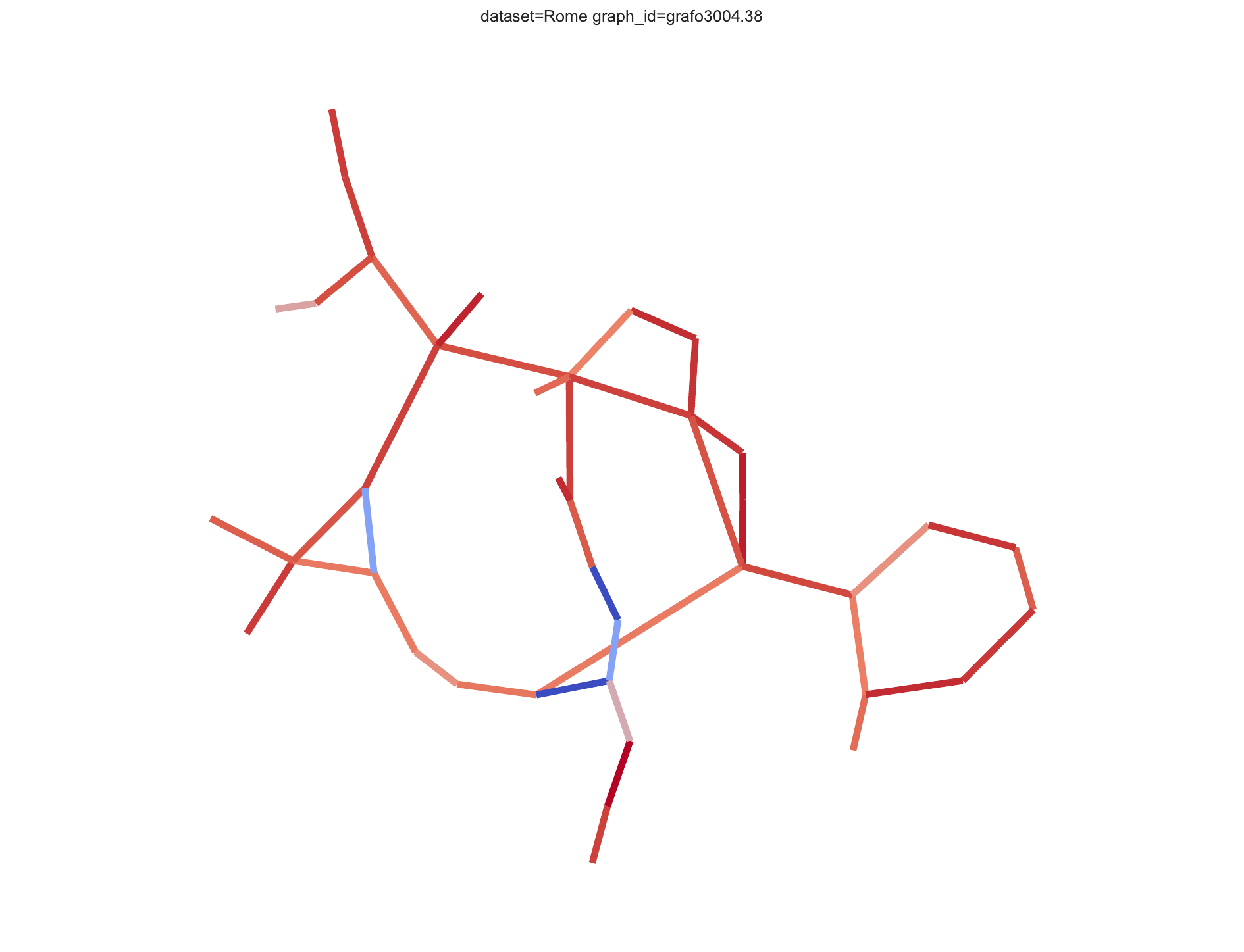} &
\imgcell{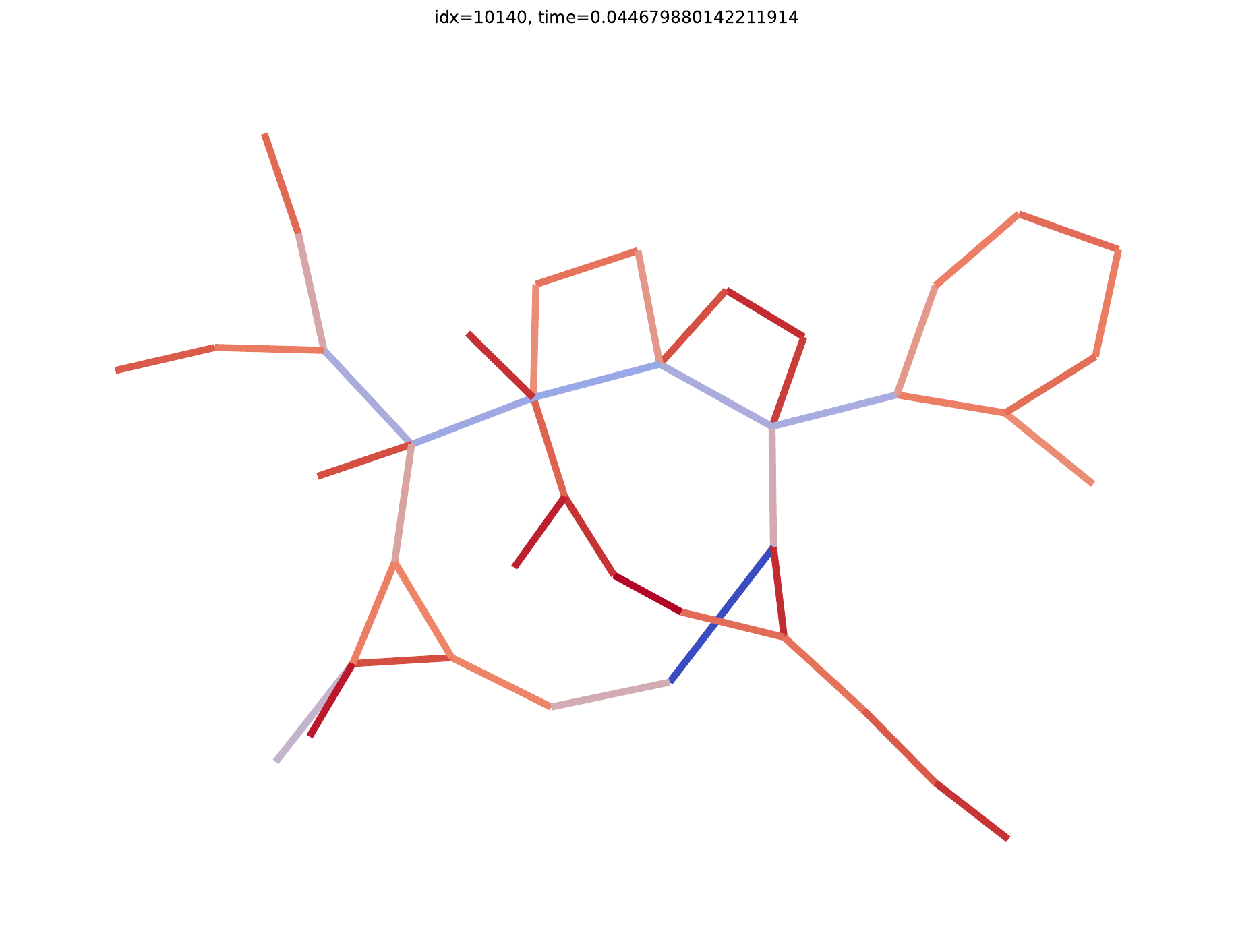} &
\imgcell{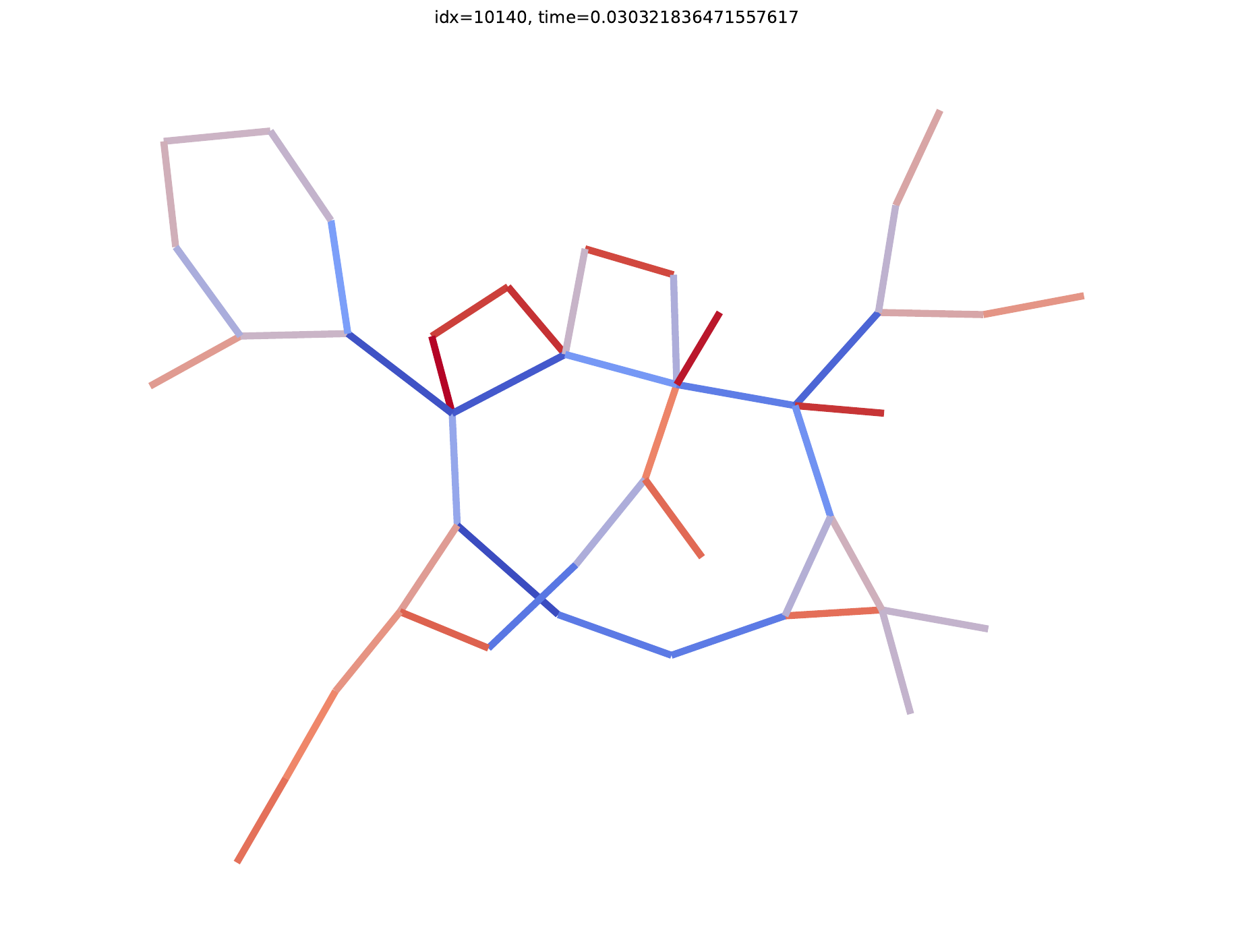} &
\imgcell{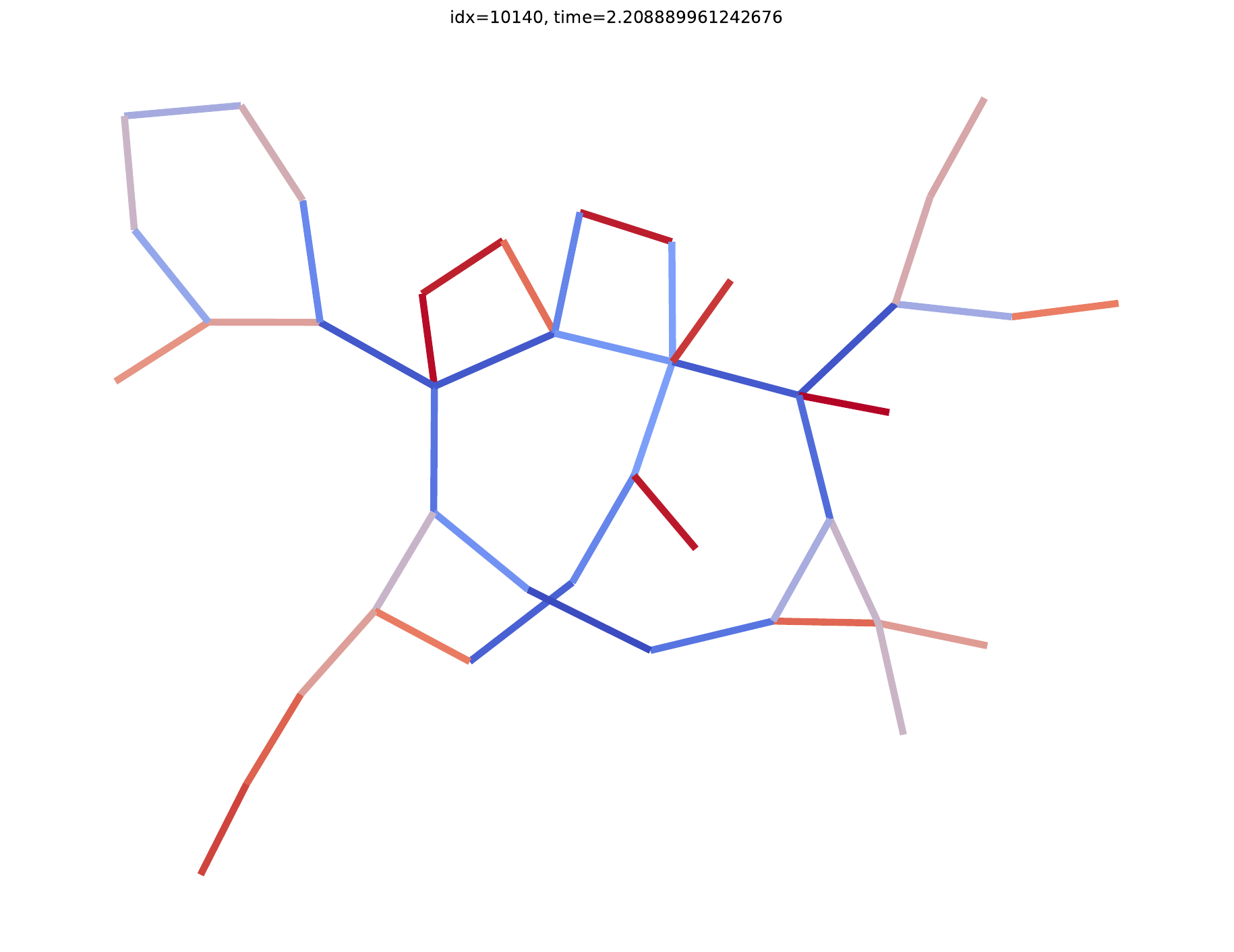} &
\imgcell{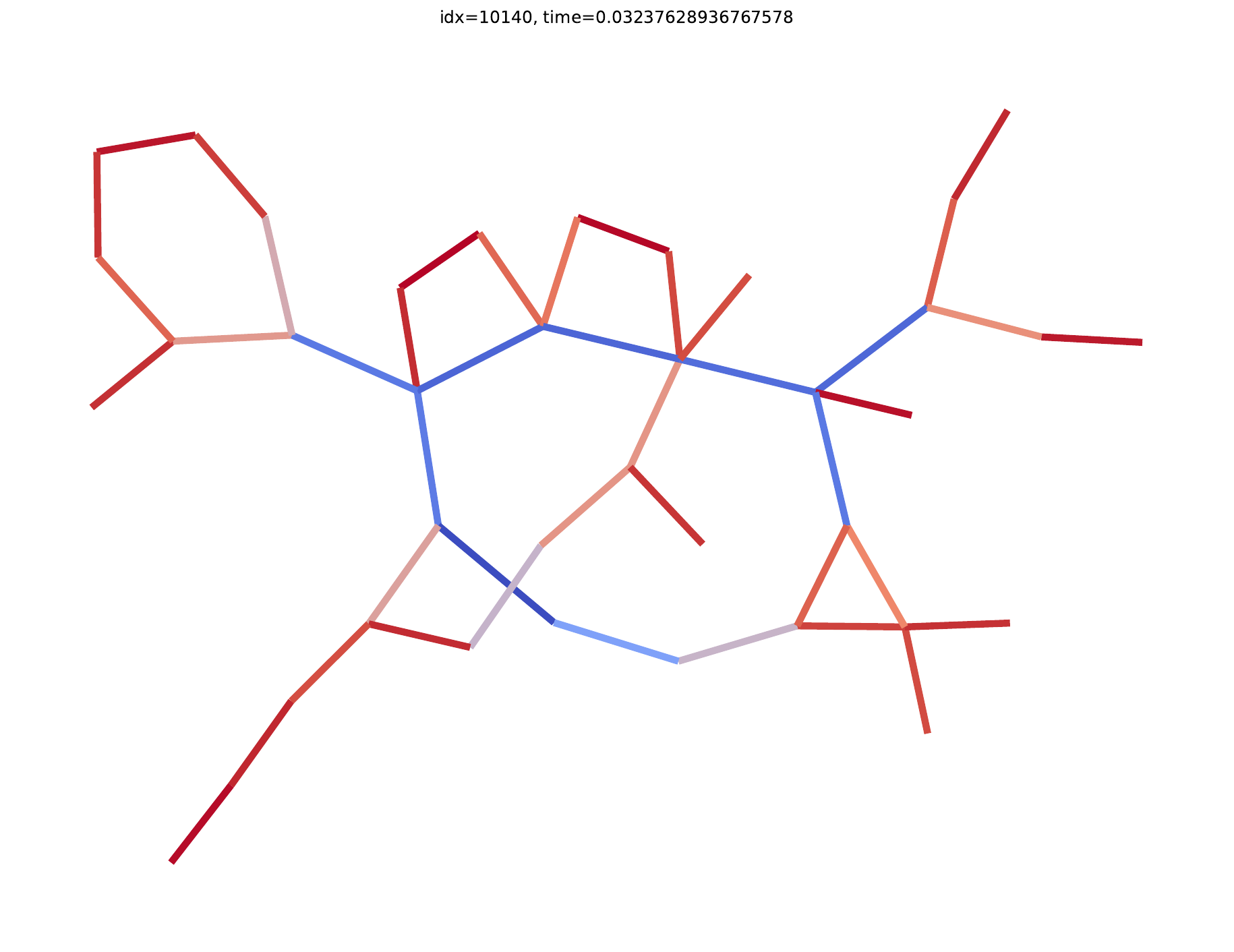} \\

&
t = 0.00s &
t = 0.33s &
t = 0.11s &
t = 0.05s &
t = 98.59s &
t = 0.05s &
t = 0.03s &
t = 0.04s &
t = 0.04s &
t = 0.03s &
t = 0.03s &
t = 0.03s \\

\makecell{\bfseries grafo9633.64\\N = 39\\M = 43} &
\imgcell{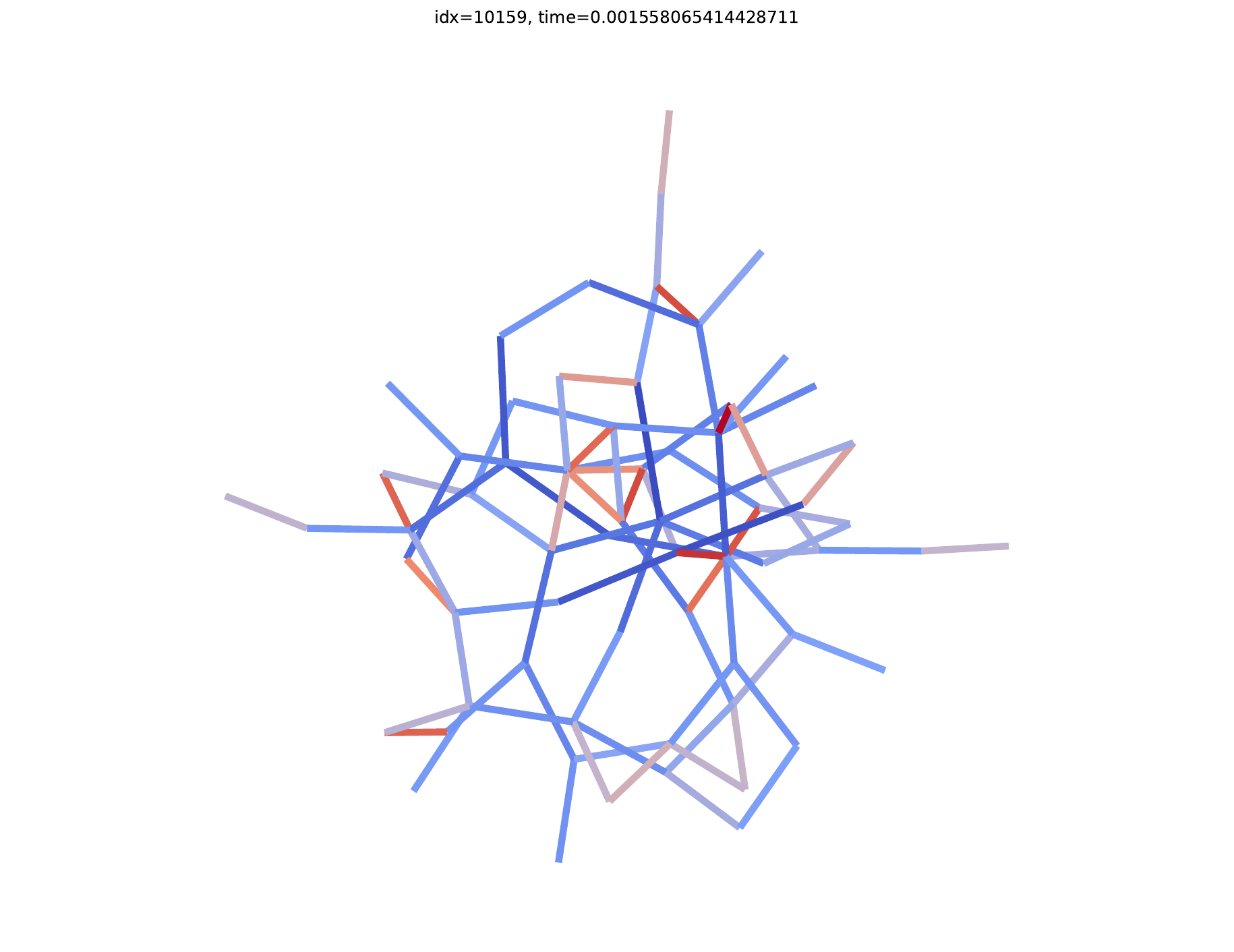} &
\imgcell{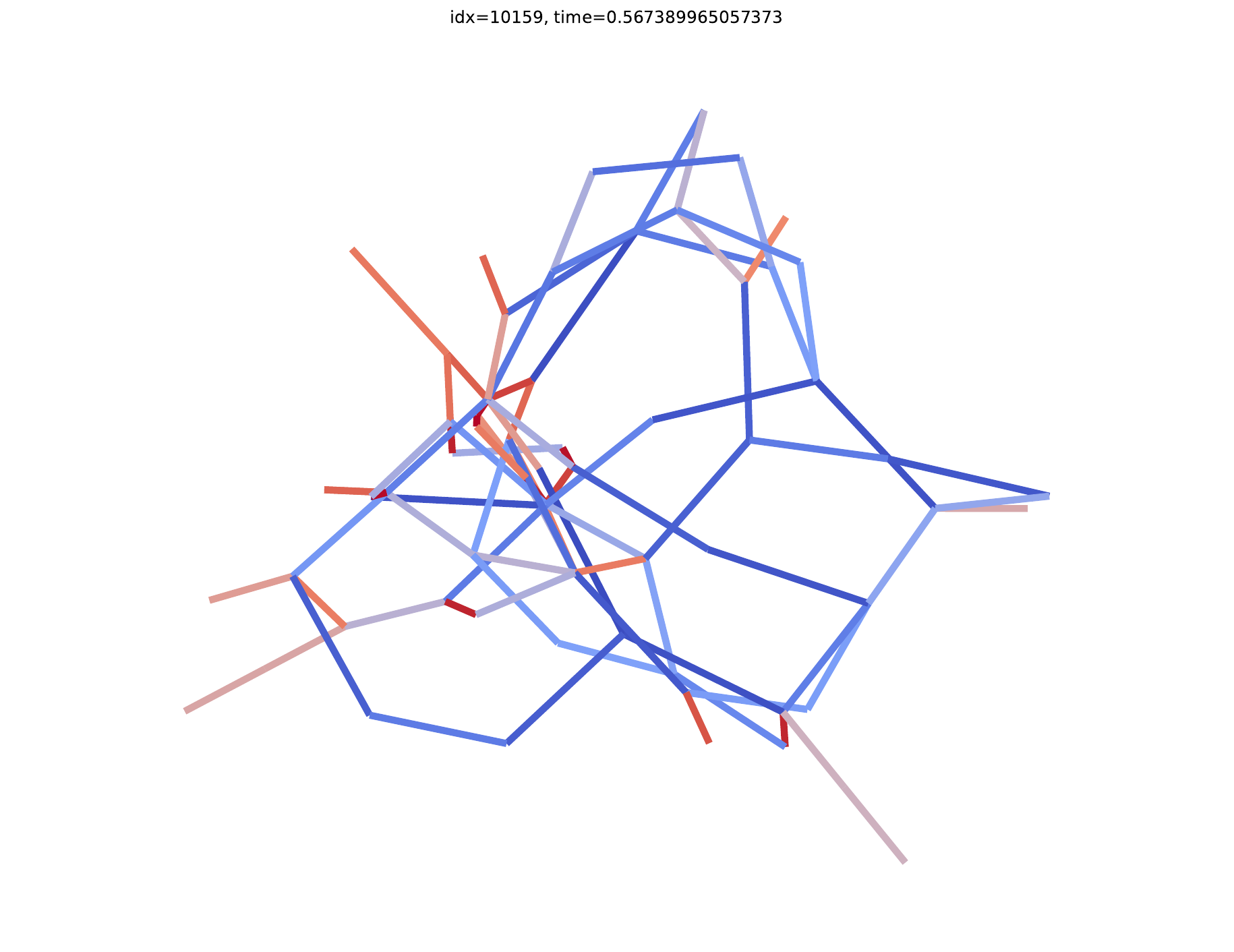} &
\imgcell{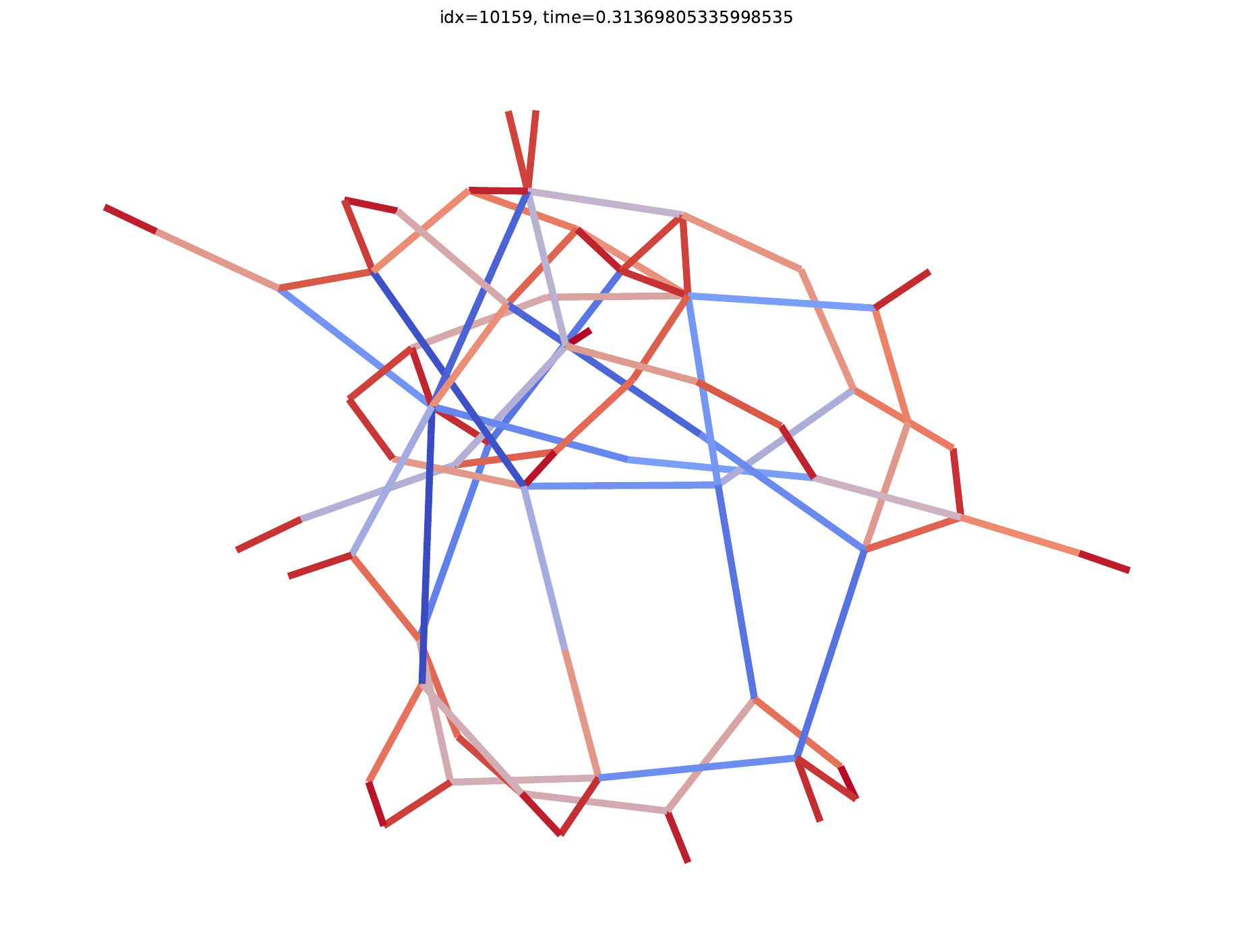} &
\imgcell{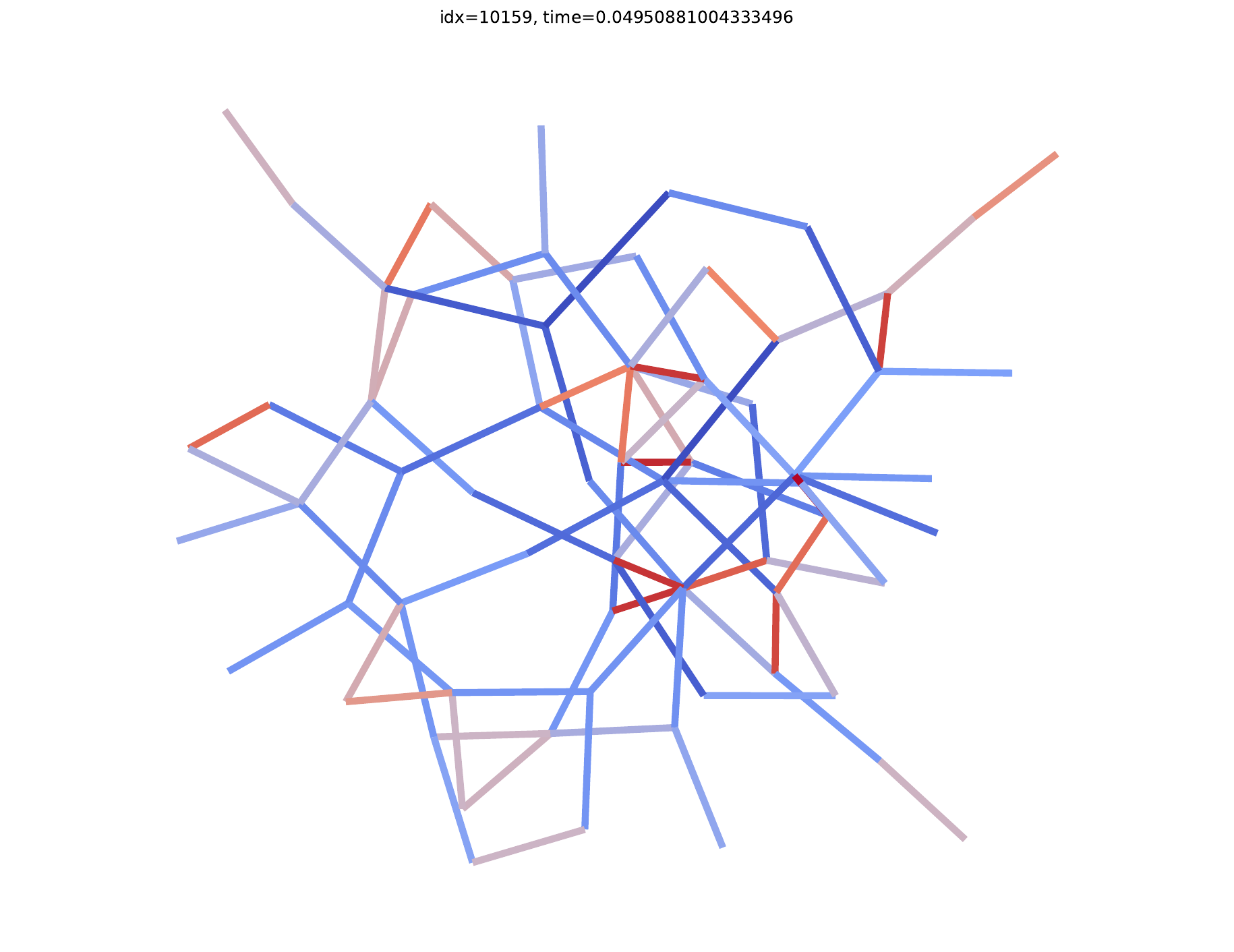} &
\imgcell{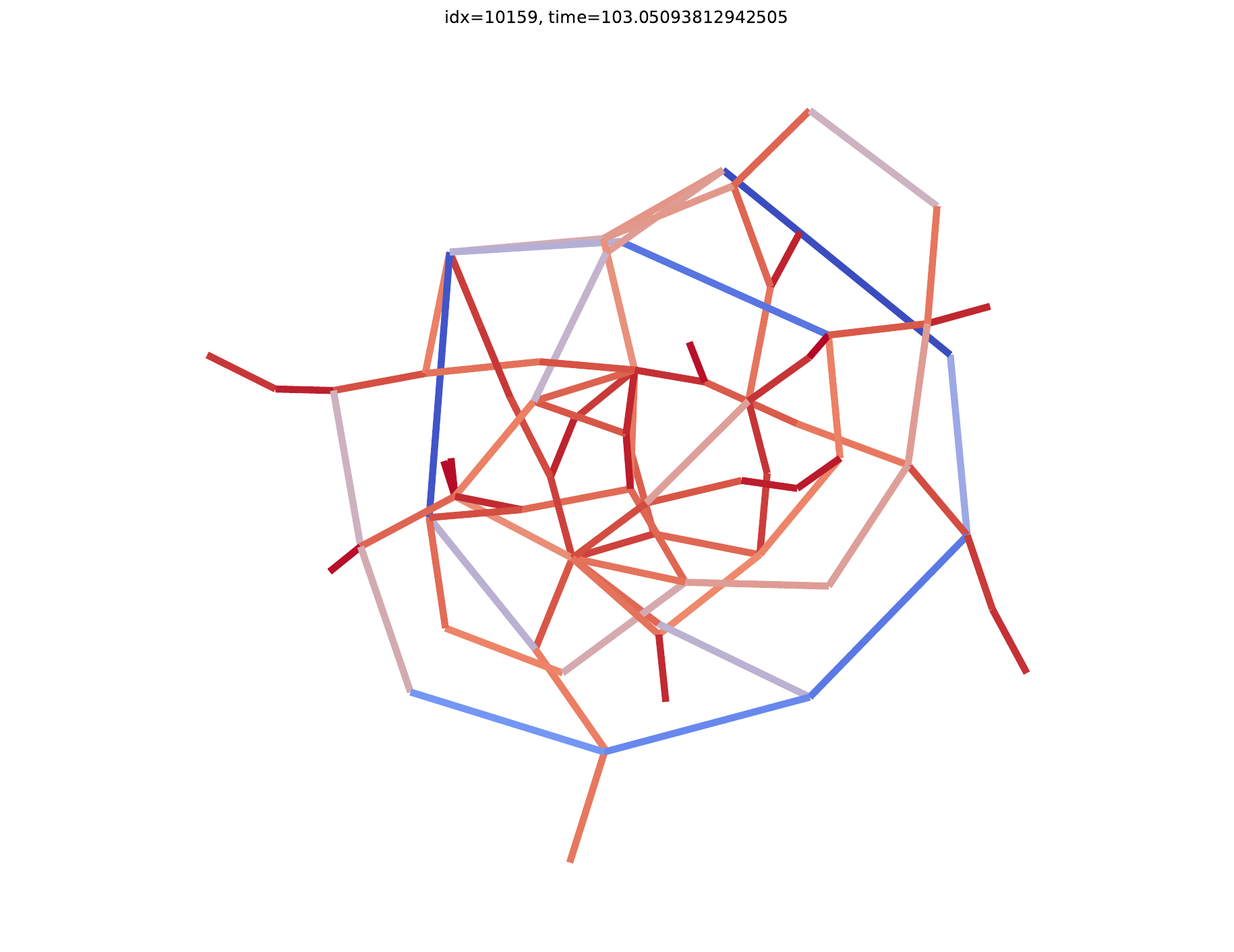} &
\imgcell{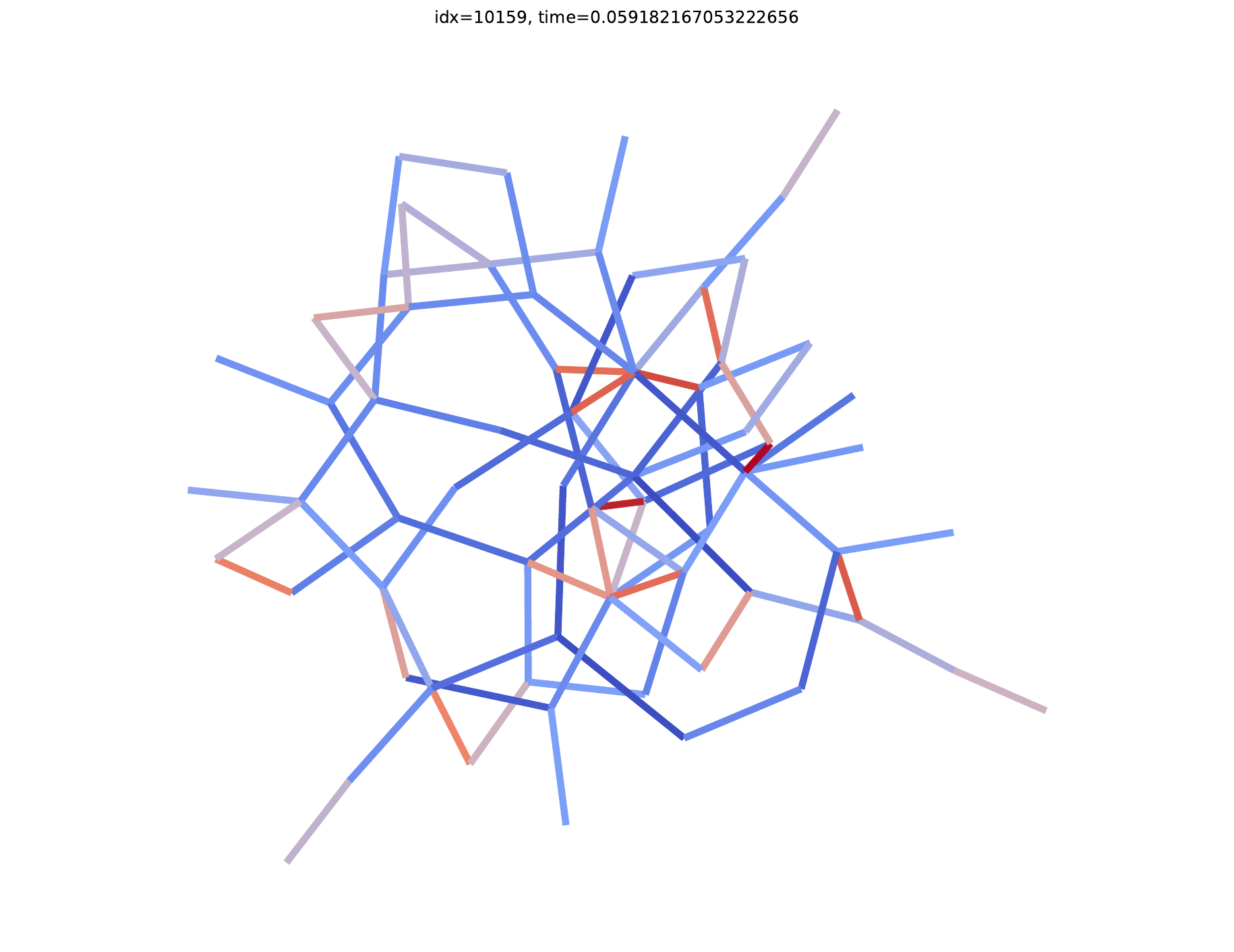} &
\imgcell{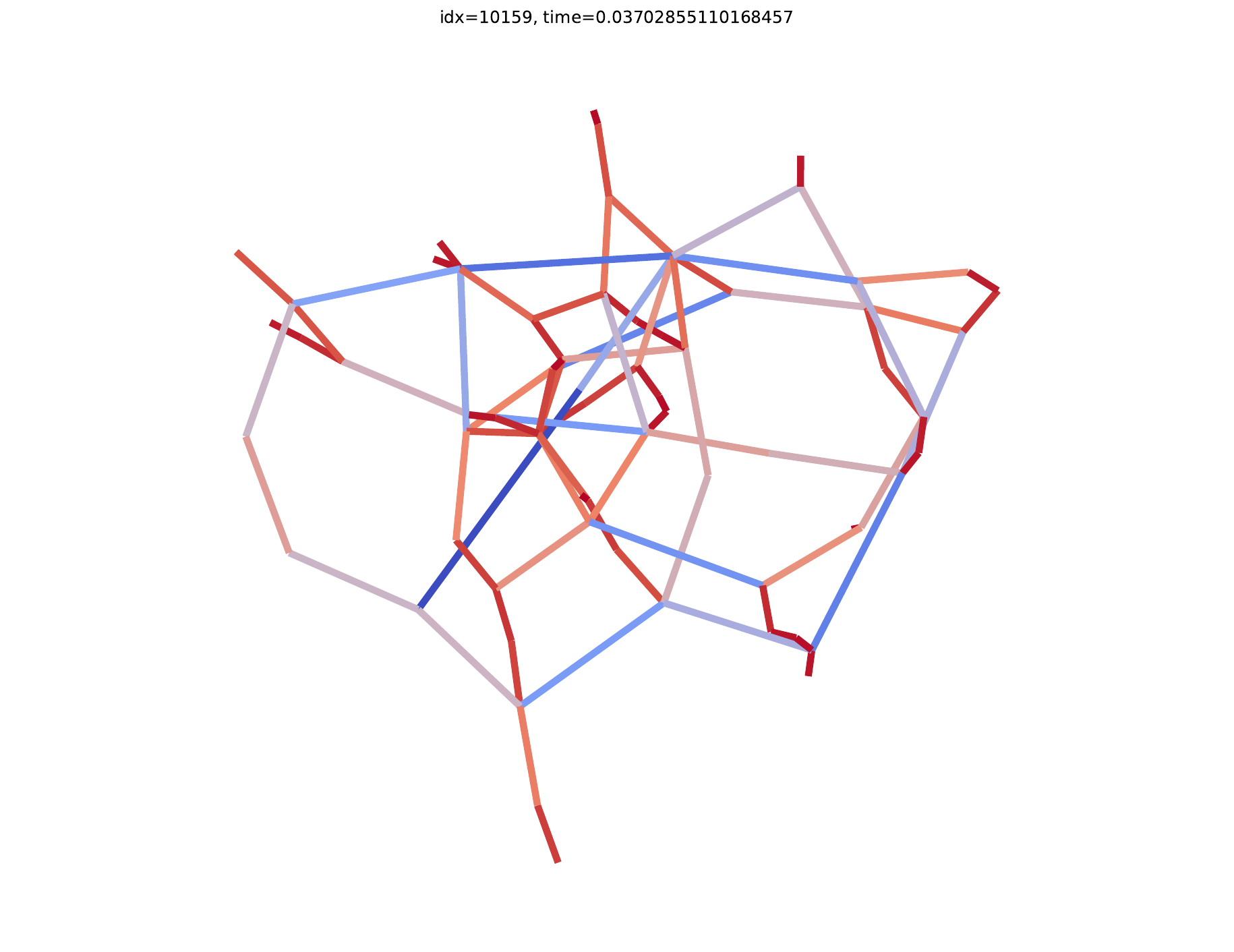} &
\imgcell{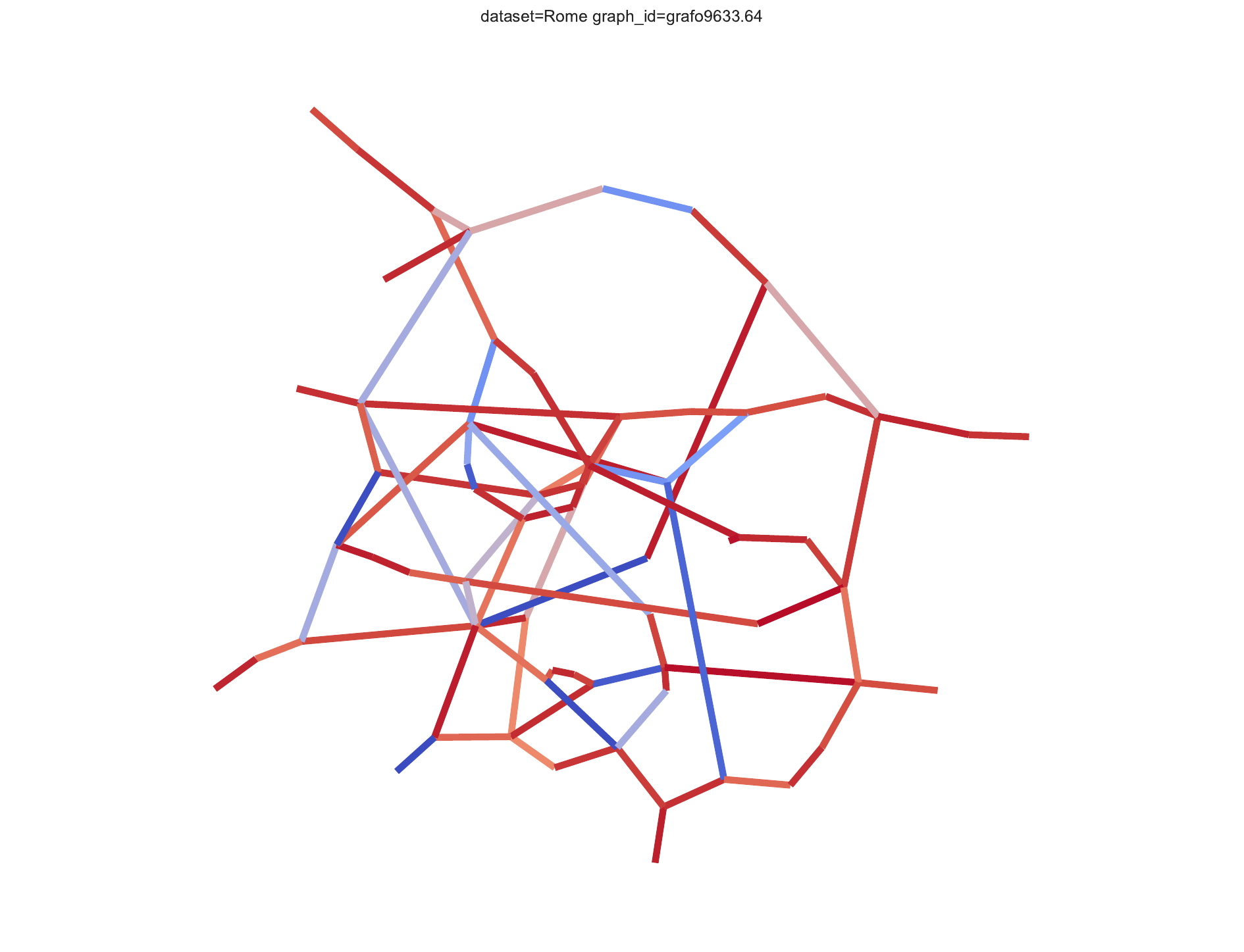} &
\imgcell{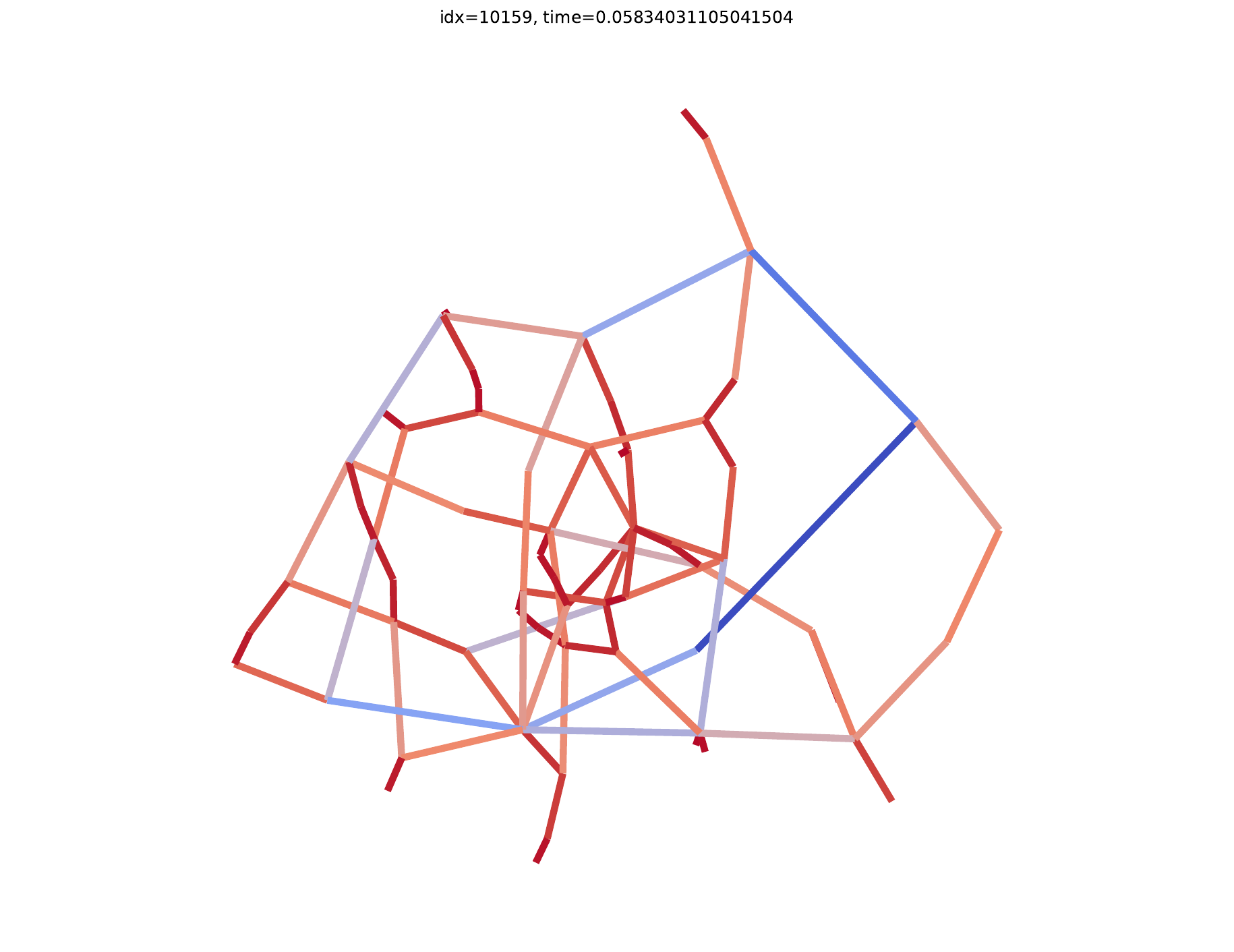} &
\imgcell{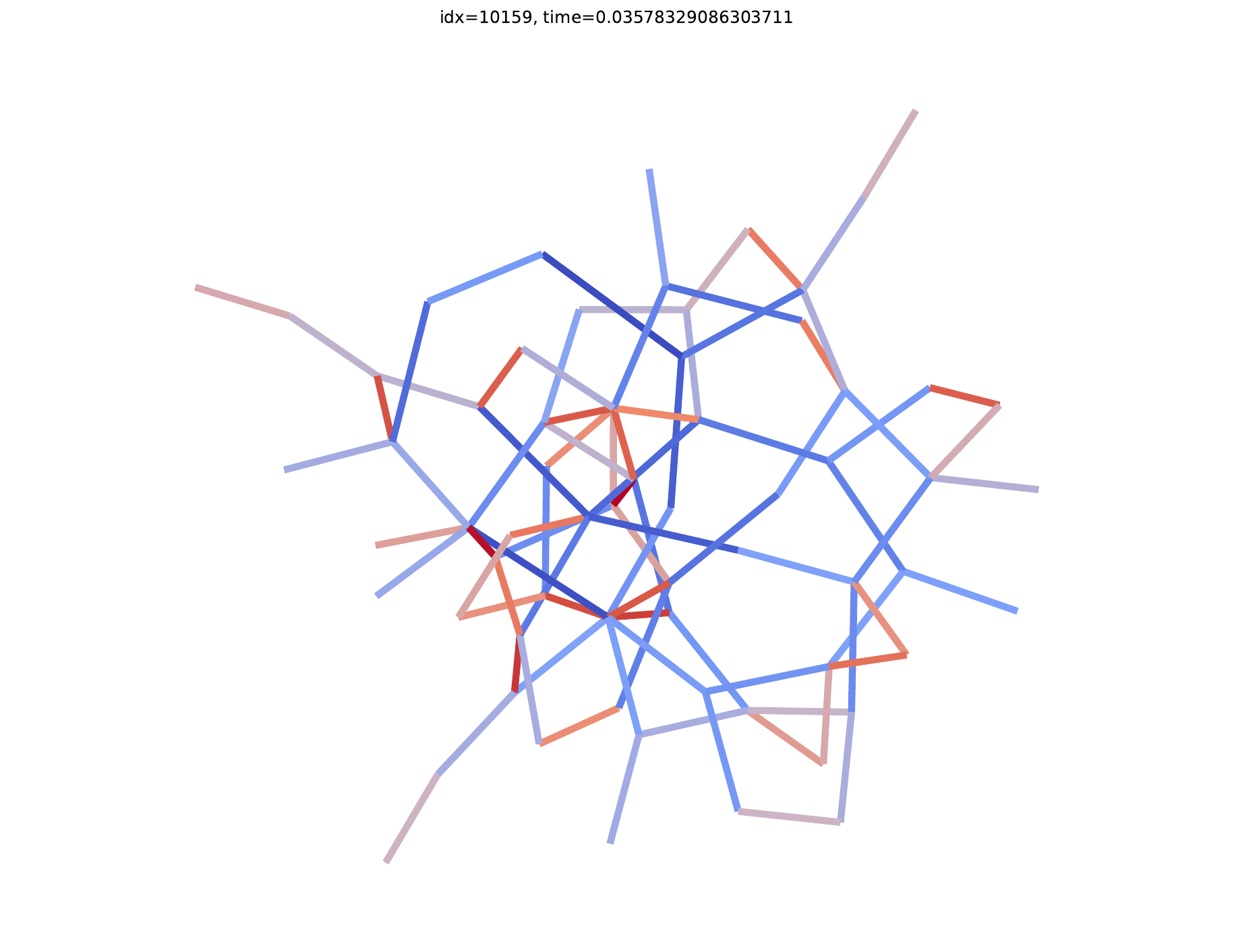} &
\imgcell{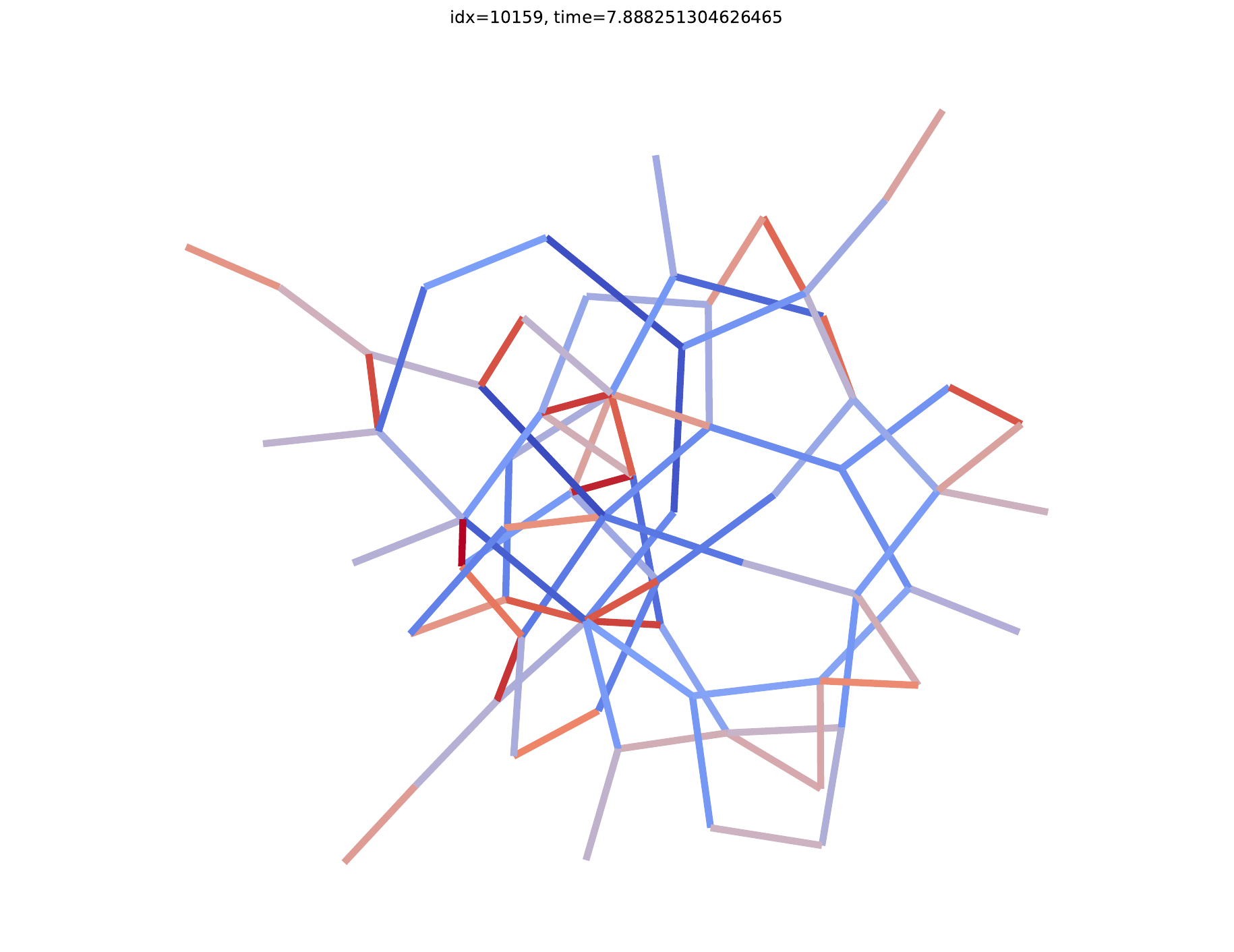} &
\imgcell{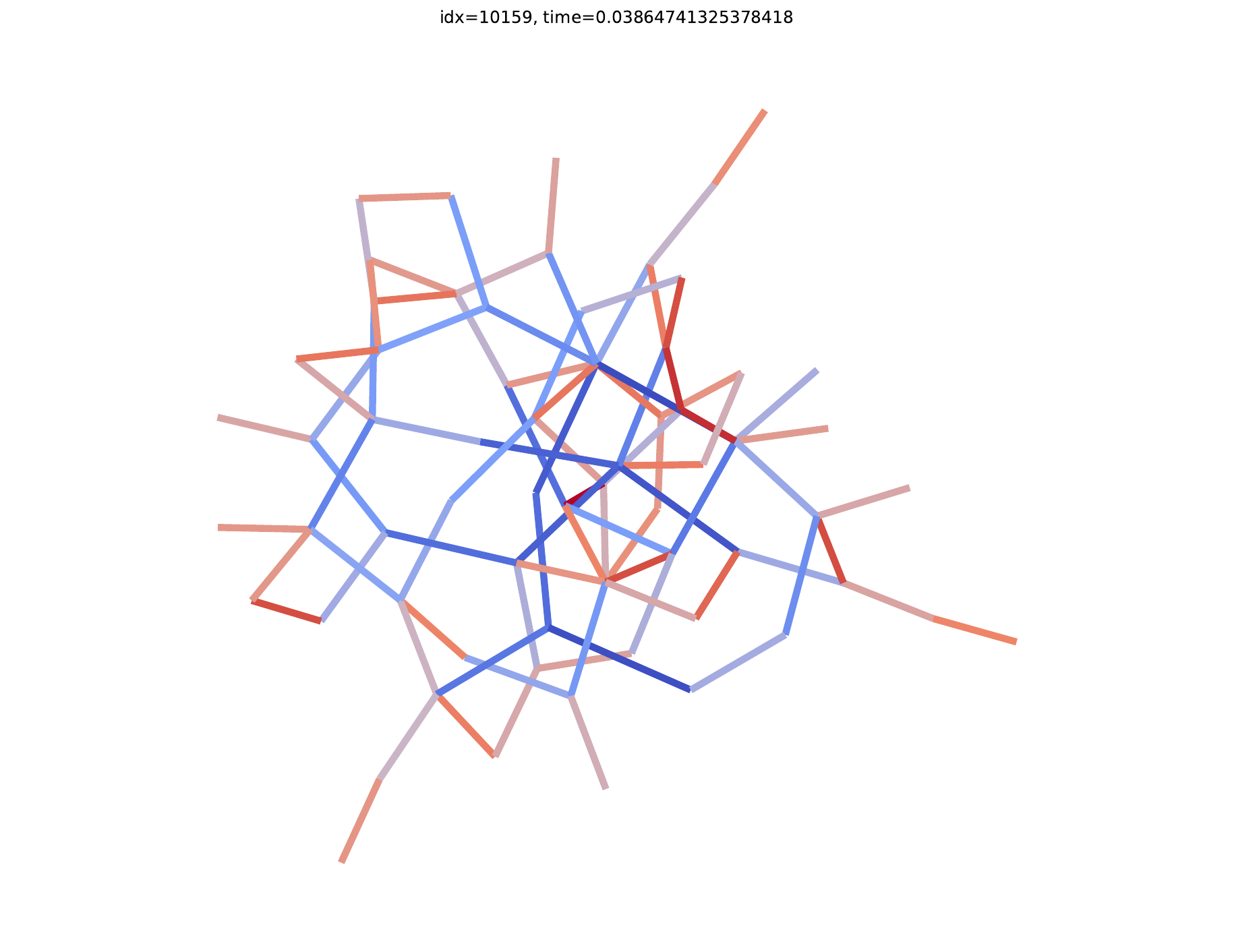} \\

&
t = 0.00s &
t = 0.57s &
t = 0.31s &
t = 0.05s &
t = 103.05s &
t = 0.06s &
t = 0.04s &
t = 0.04s &
t = 0.06s &
t = 0.04s &
t = 0.05s &
t = 0.04s \\

\makecell{\bfseries grafo4551.61\\N = 85\\M = 108} &
\imgcell{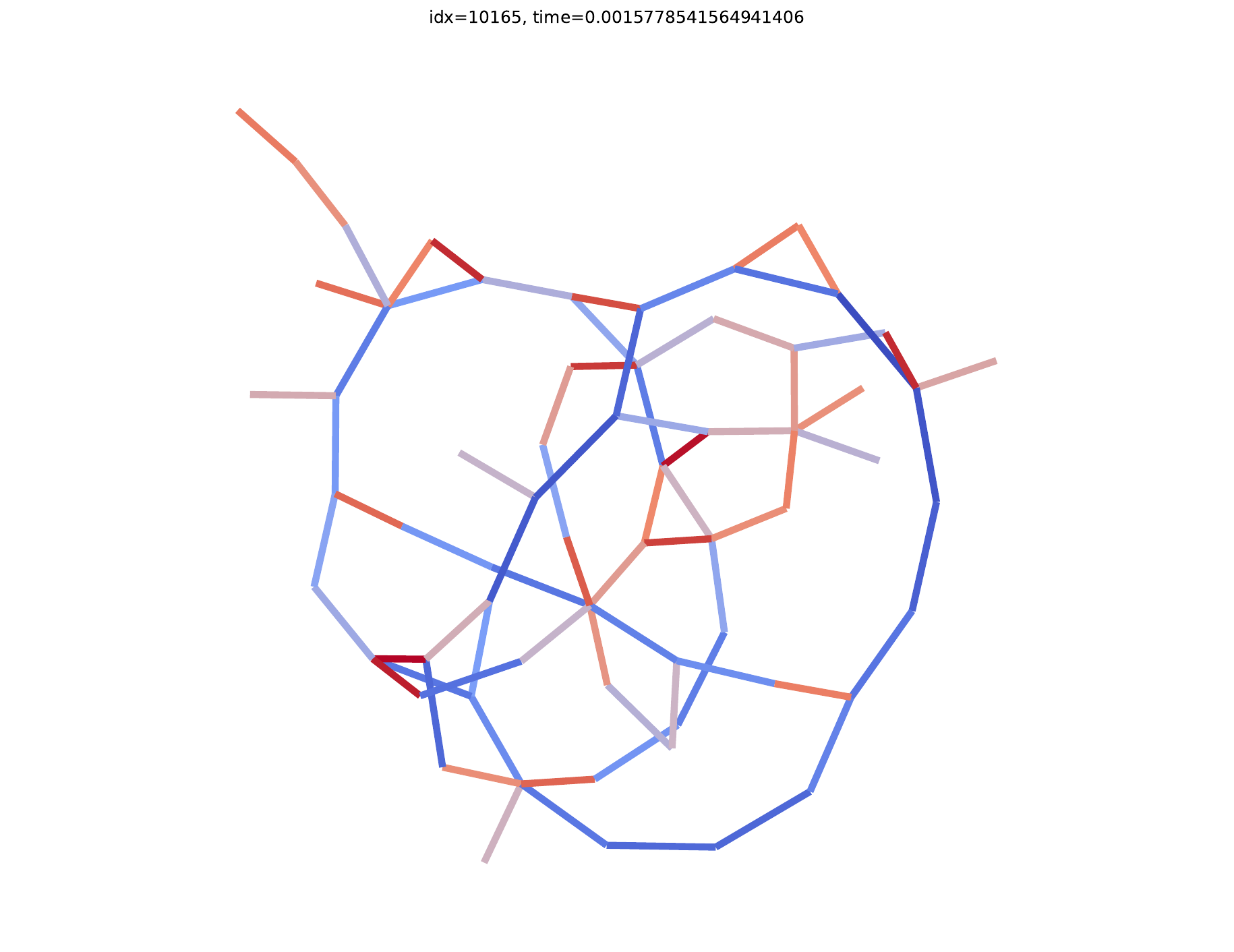} &
\imgcell{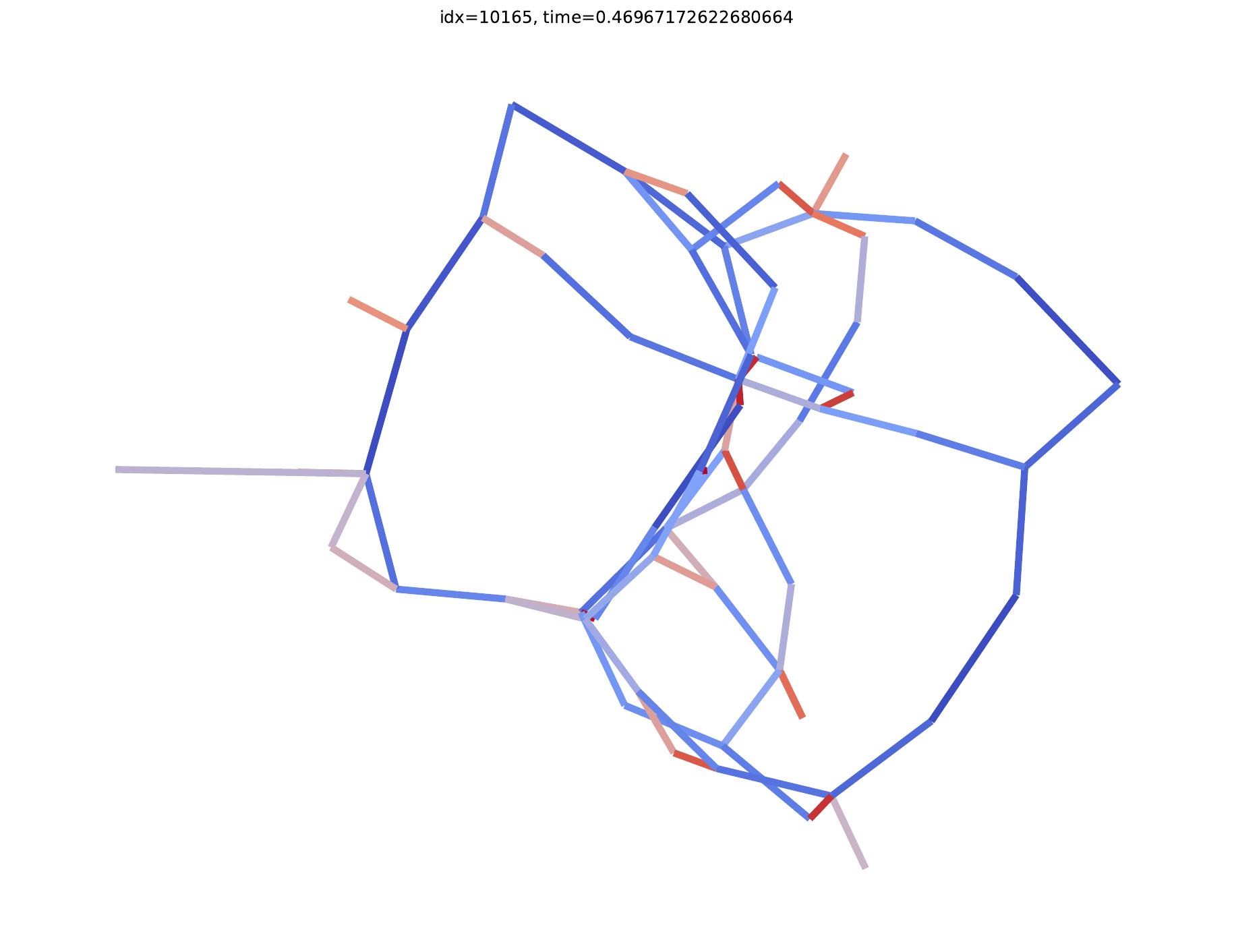} &
\imgcell{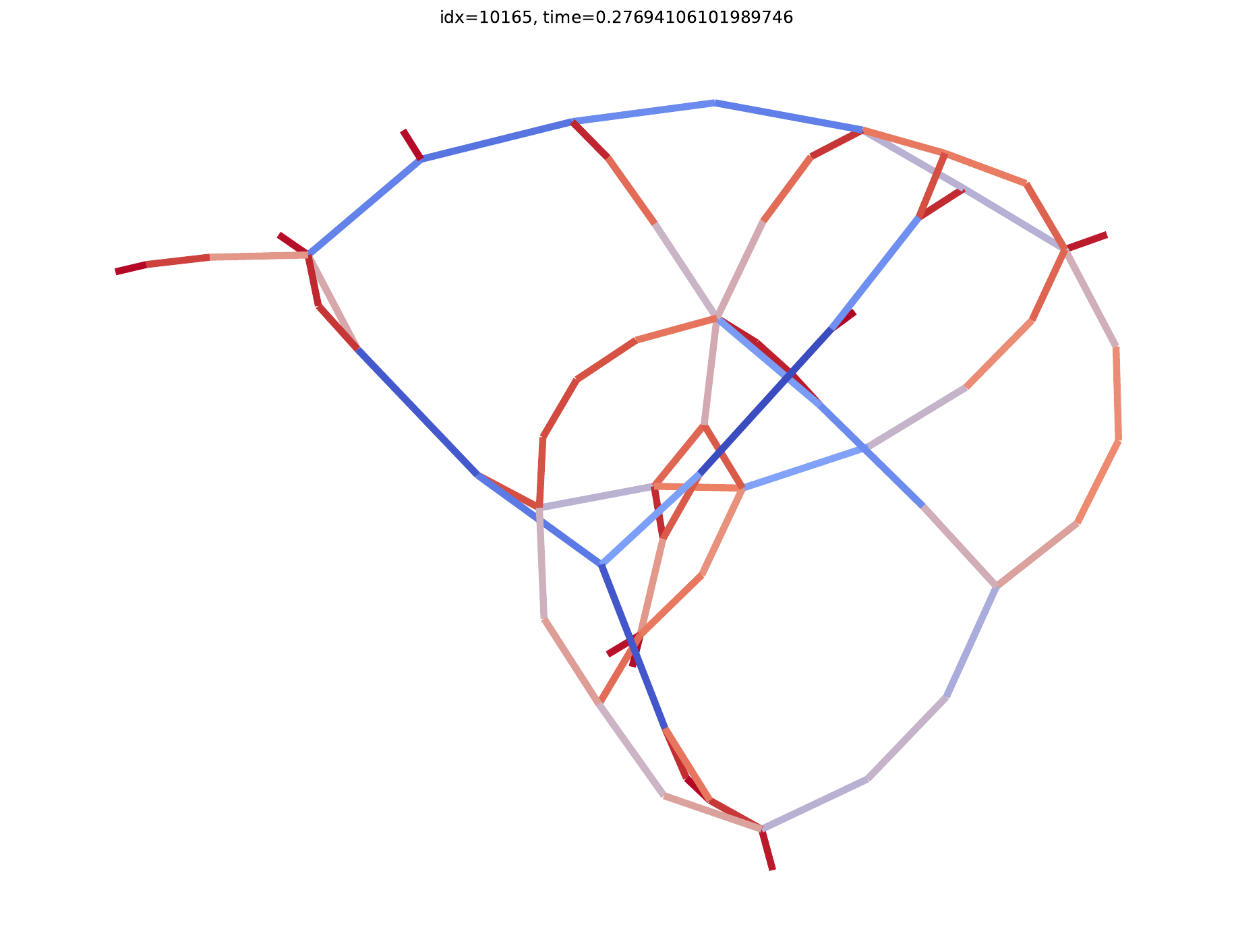} &
\imgcell{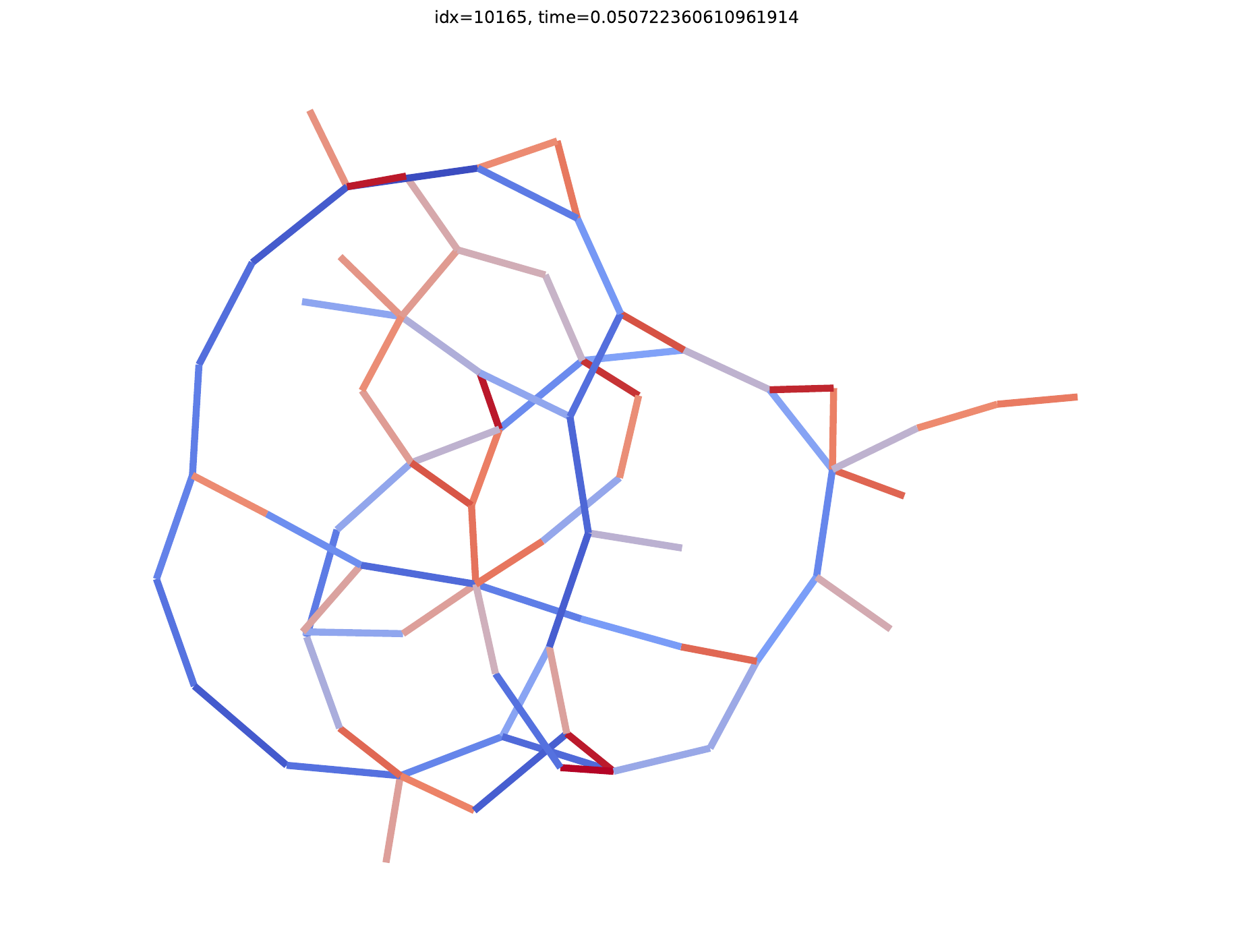} &
\imgcell{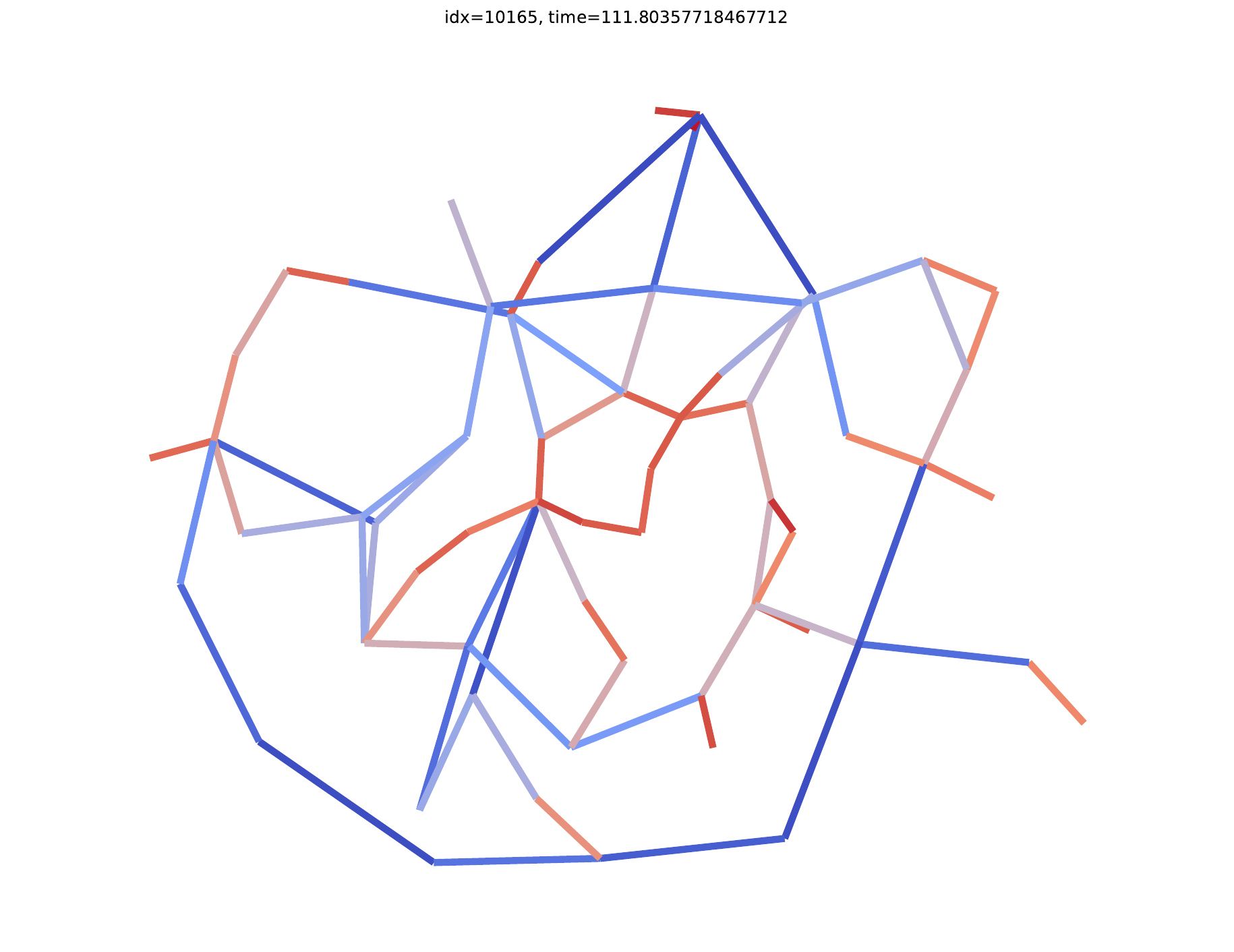} &
\imgcell{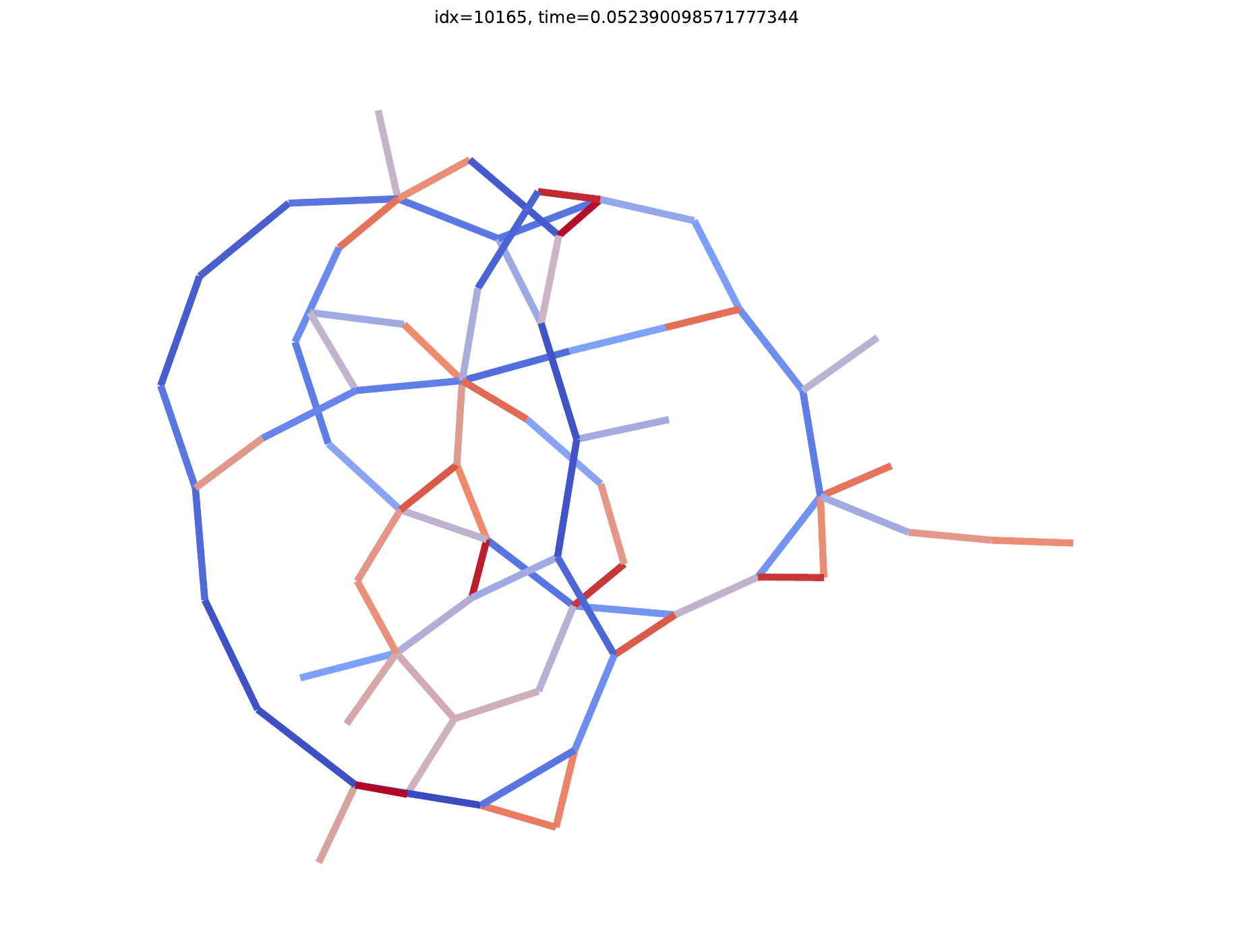} &
\imgcell{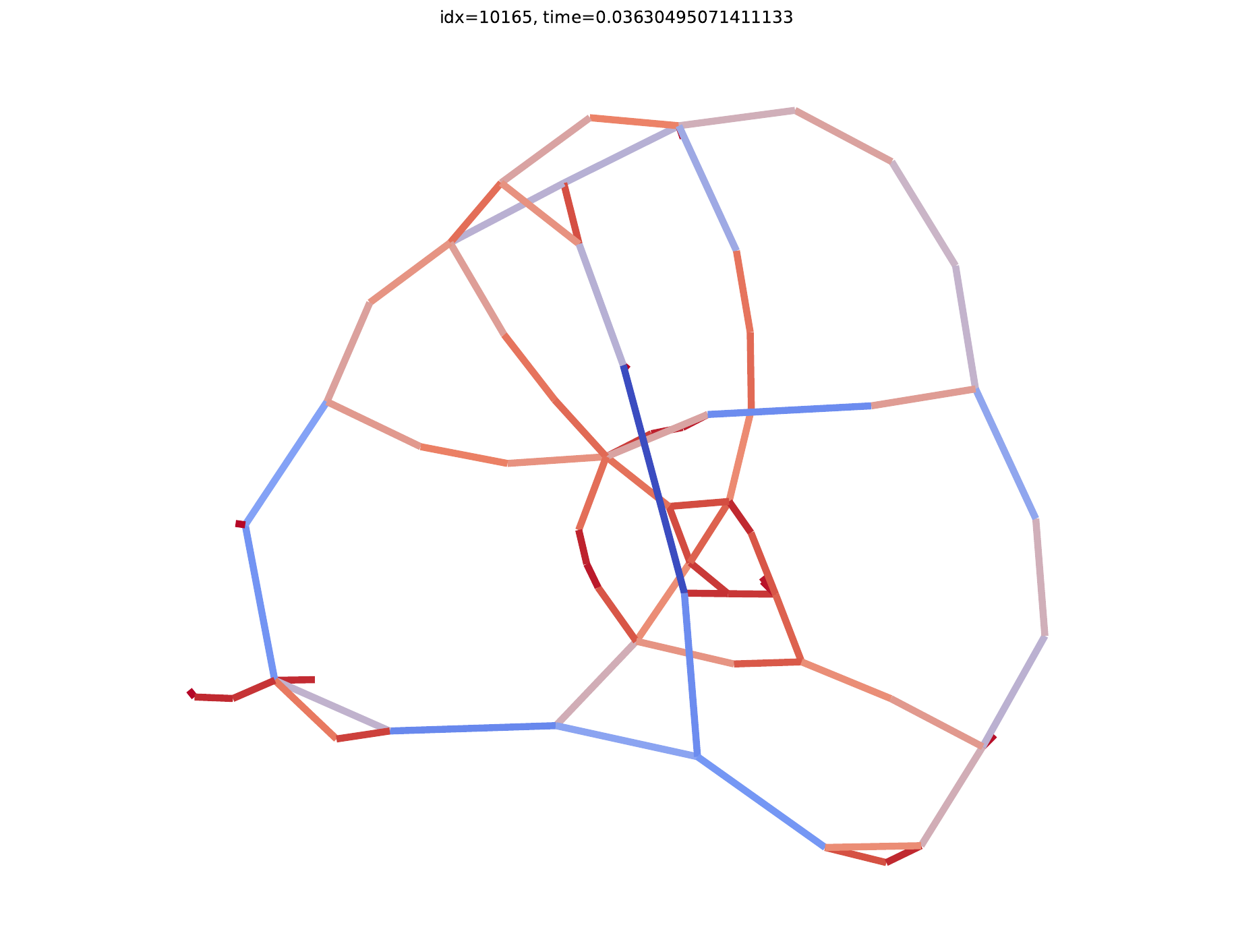} &
\imgcell{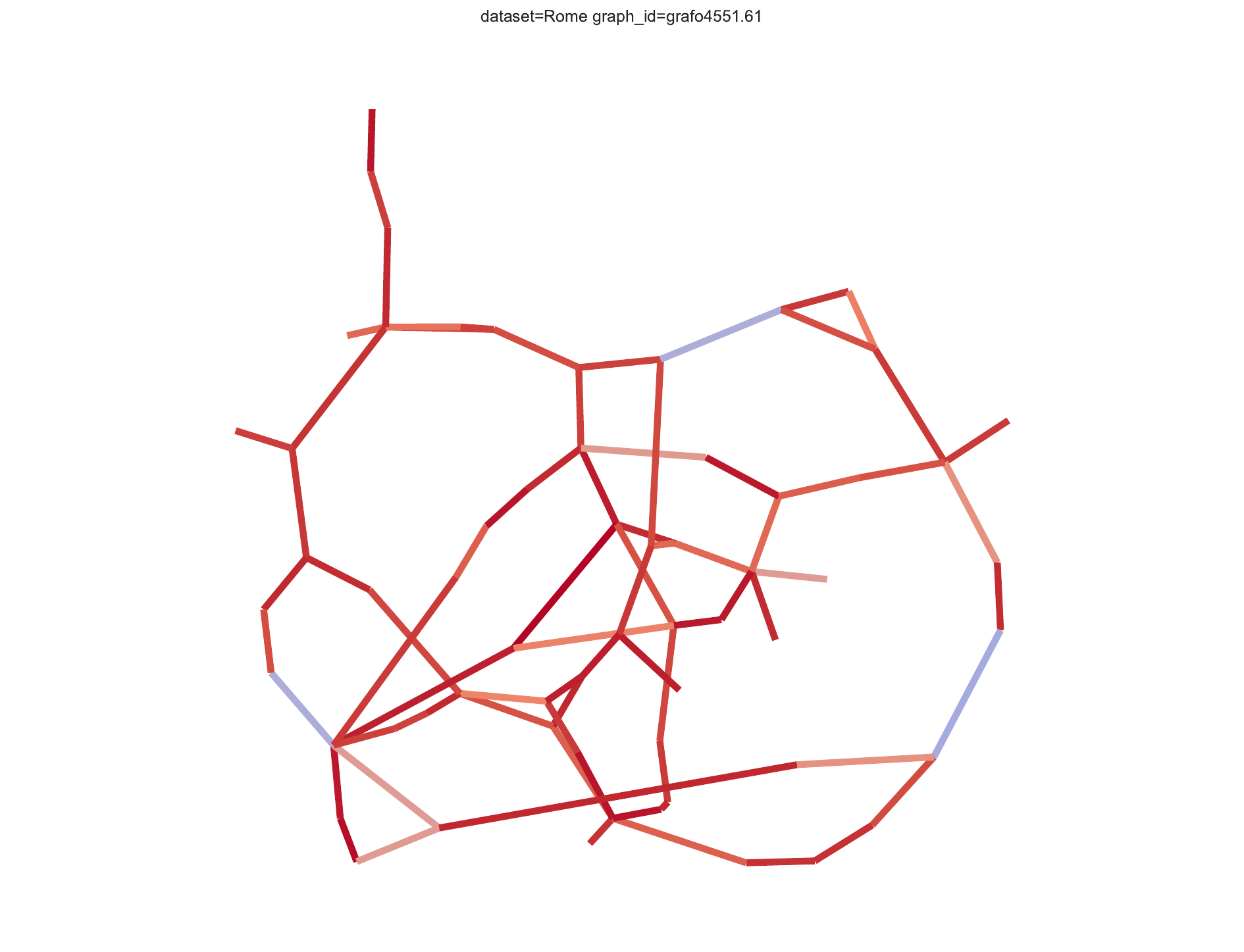} &
\imgcell{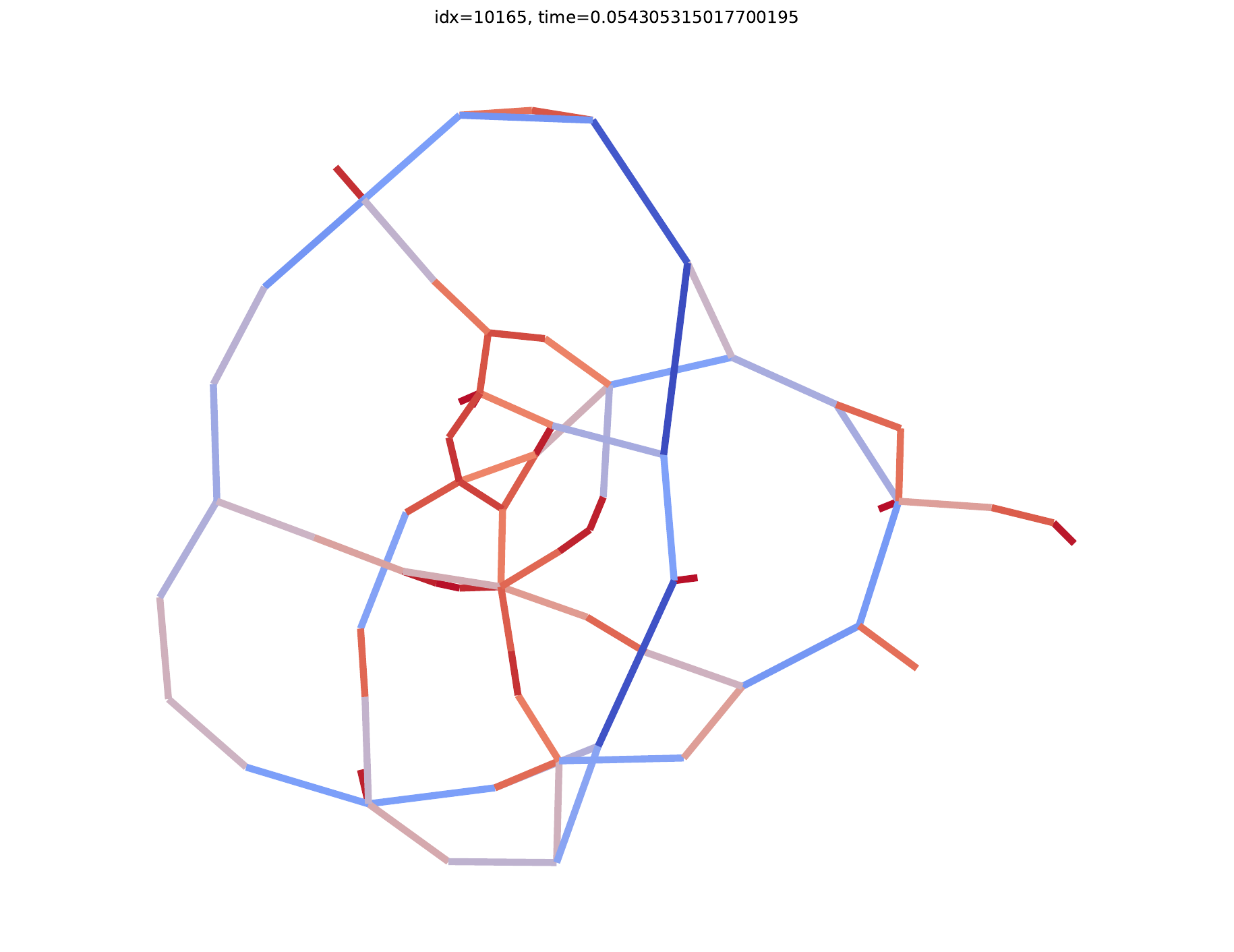} &
\imgcell{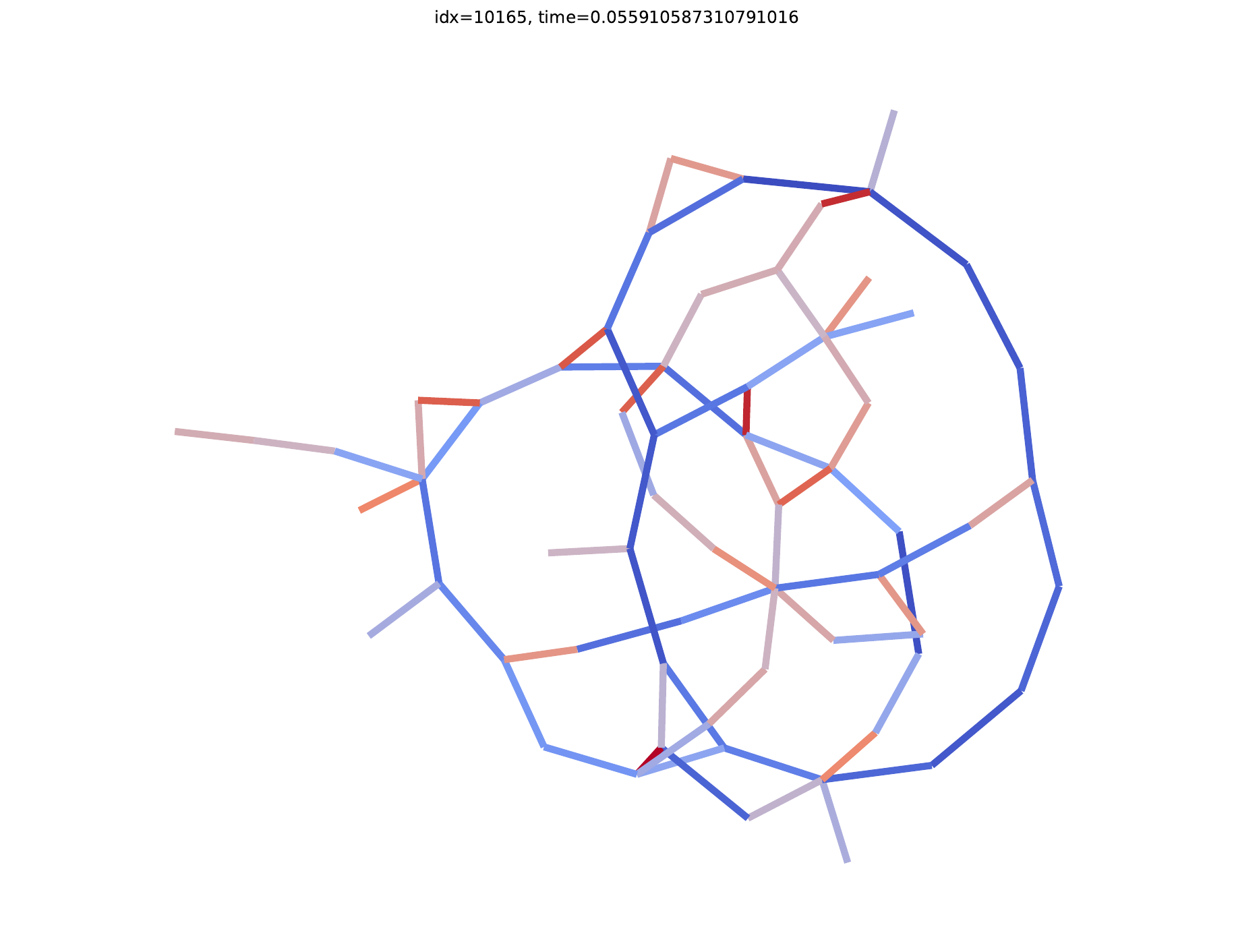} &
\imgcell{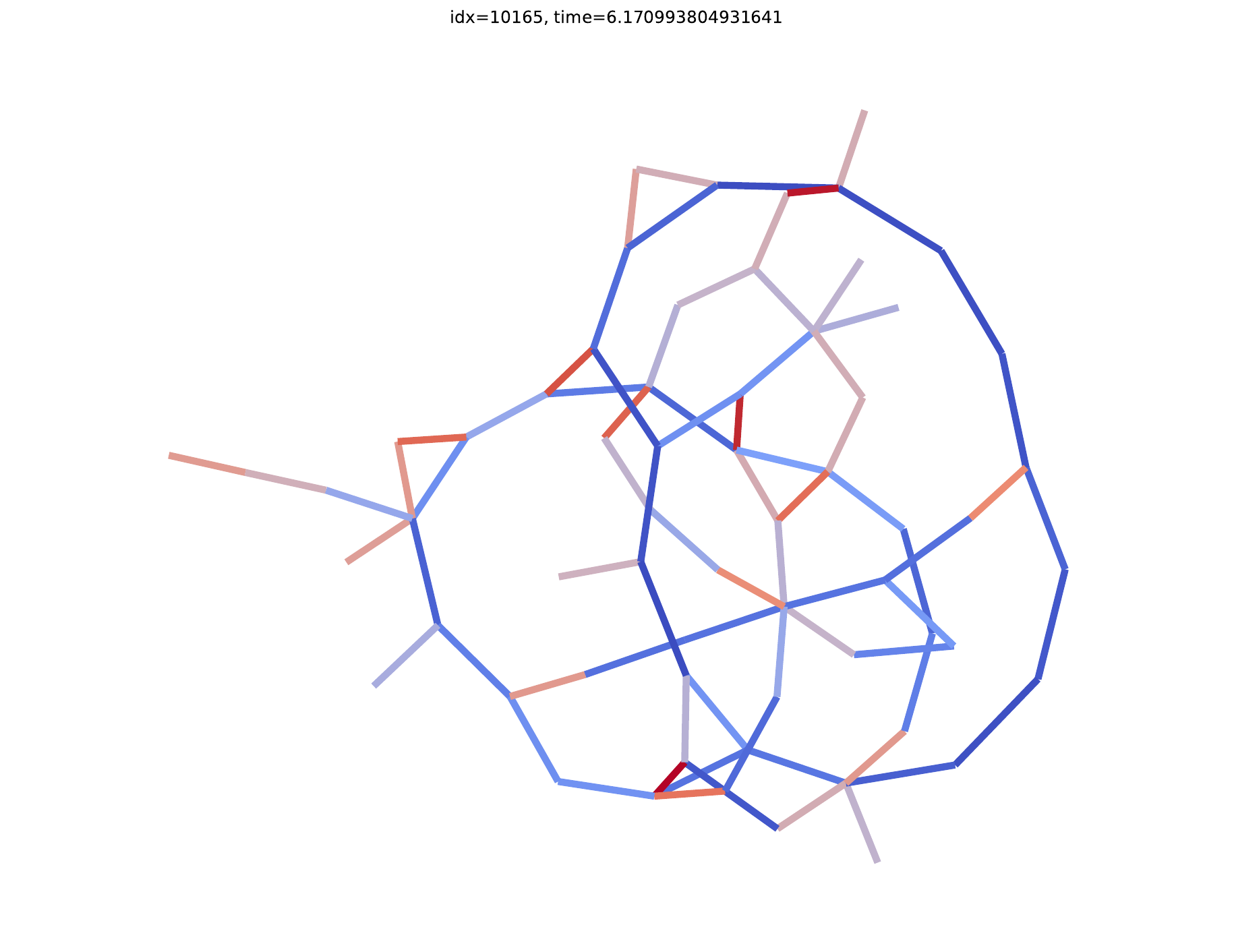} &
\imgcell{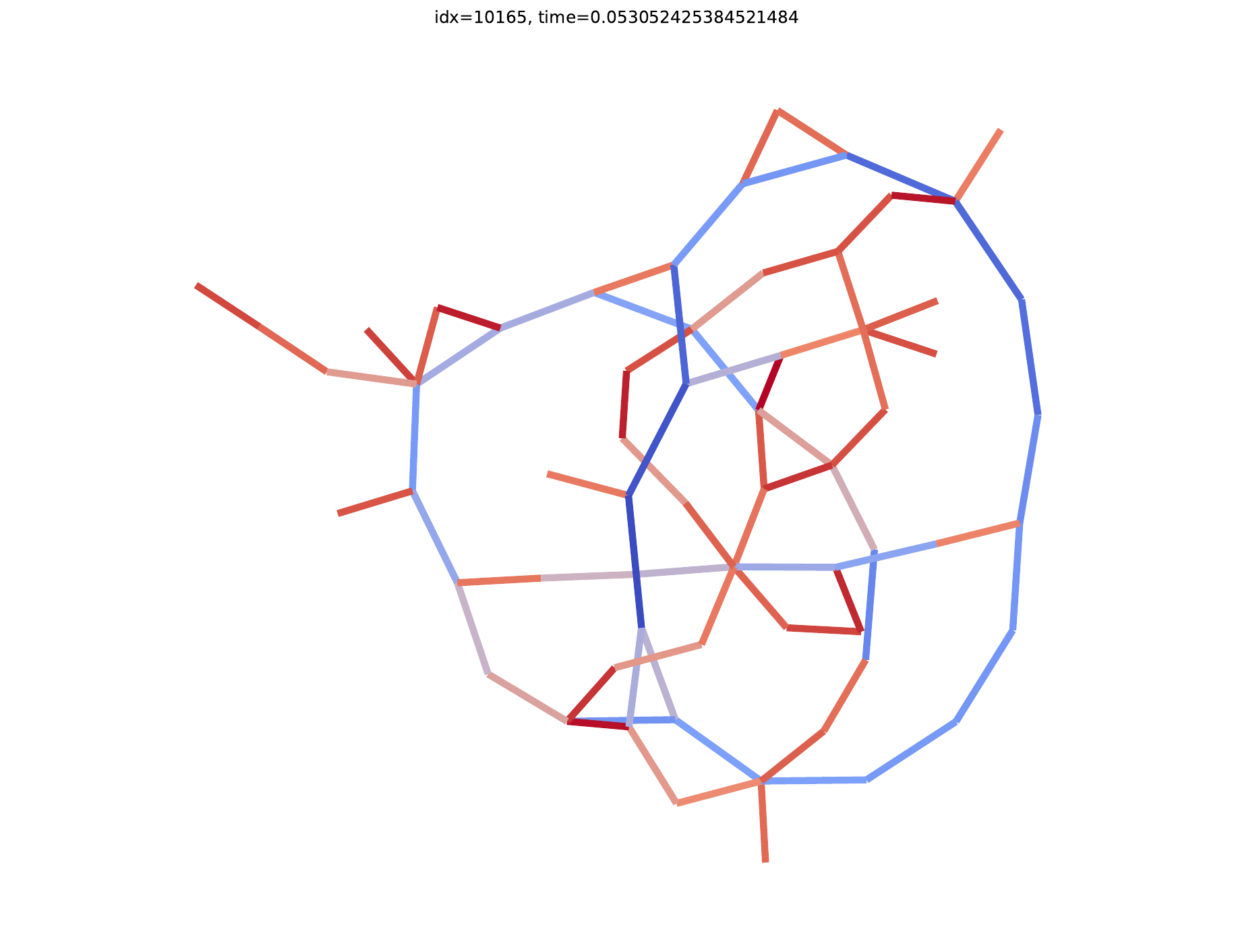} \\

&
t = 0.00s &
t = 0.47s &
t = 0.28s &
t = 0.05s &
t = 111.80s &
t = 0.05s &
t = 0.04s &
t = 0.05s &
t = 0.05s &
t = 0.06s &
t = 0.07s &
t = 0.05s \\

\makecell{\bfseries grafo10802.37\\N = 94\\M = 122} &
\imgcell{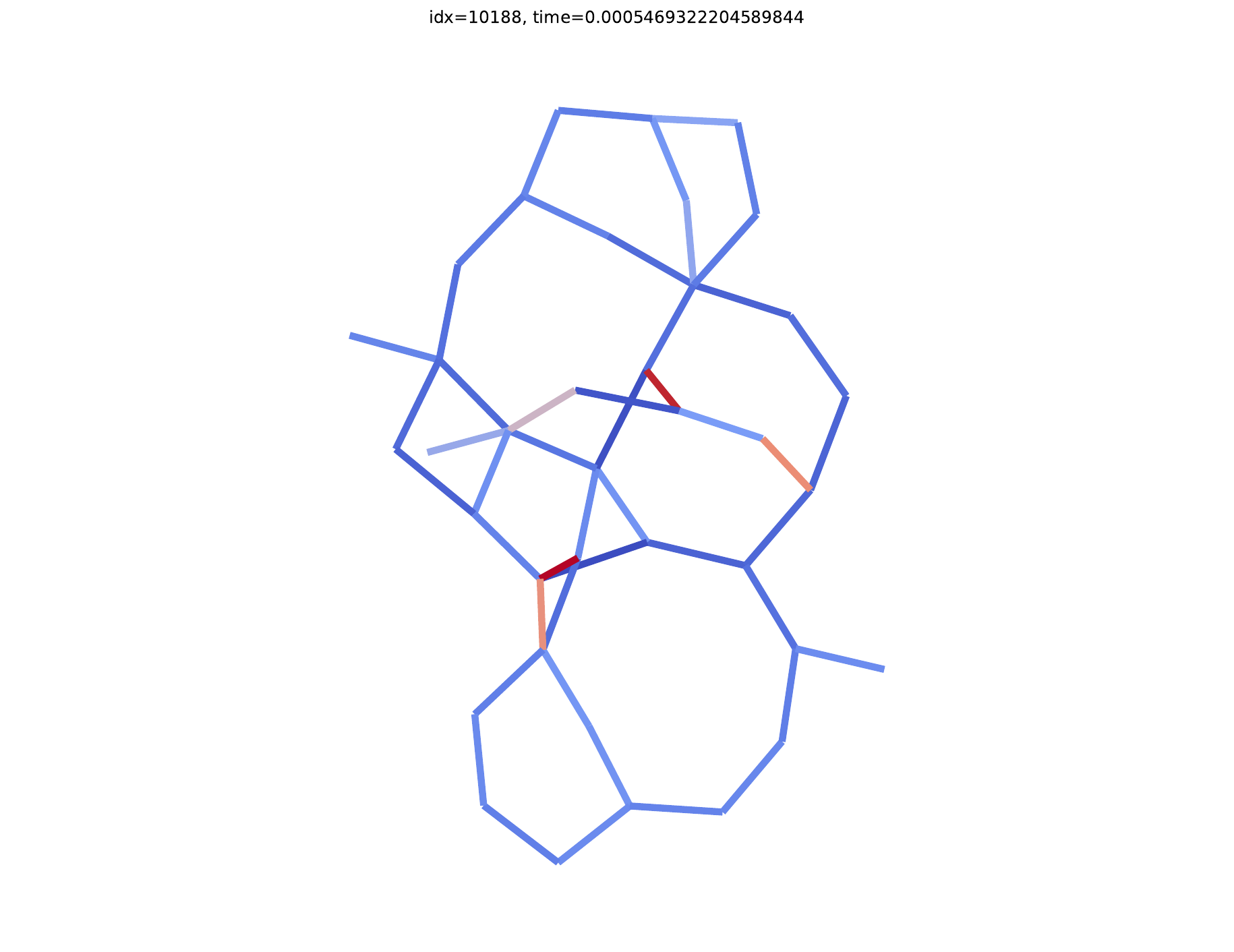} &
\imgcell{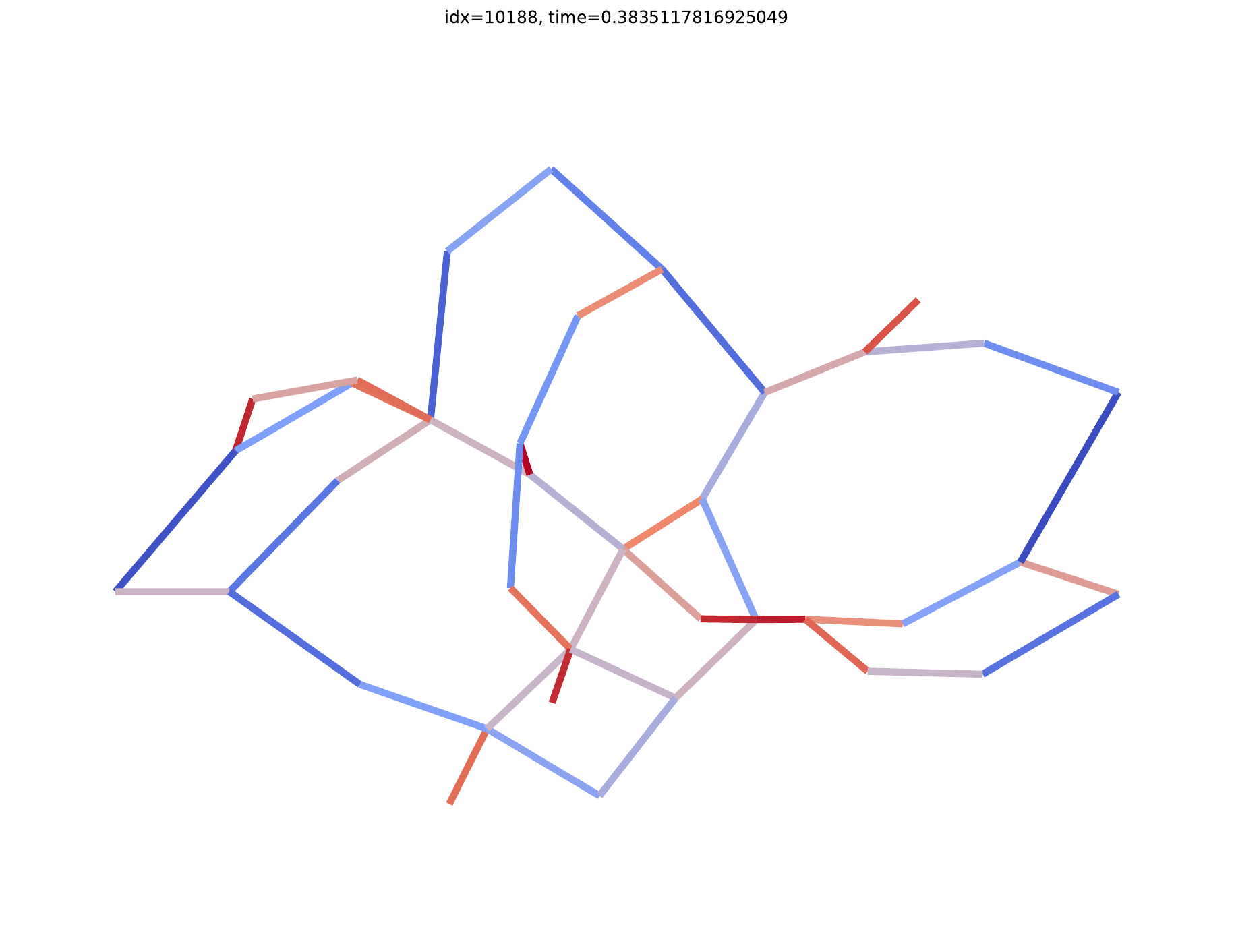} &
\imgcell{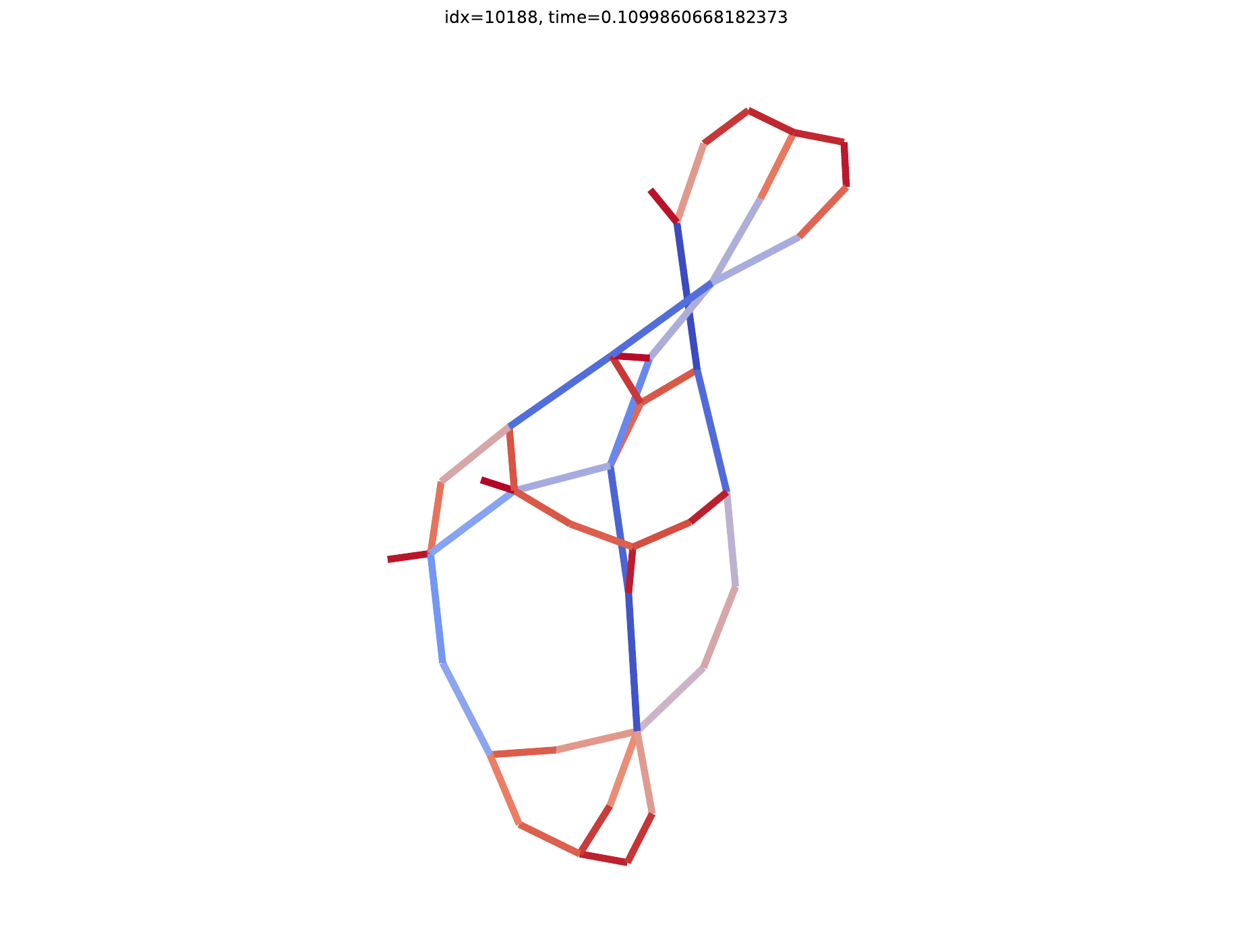} &
\imgcell{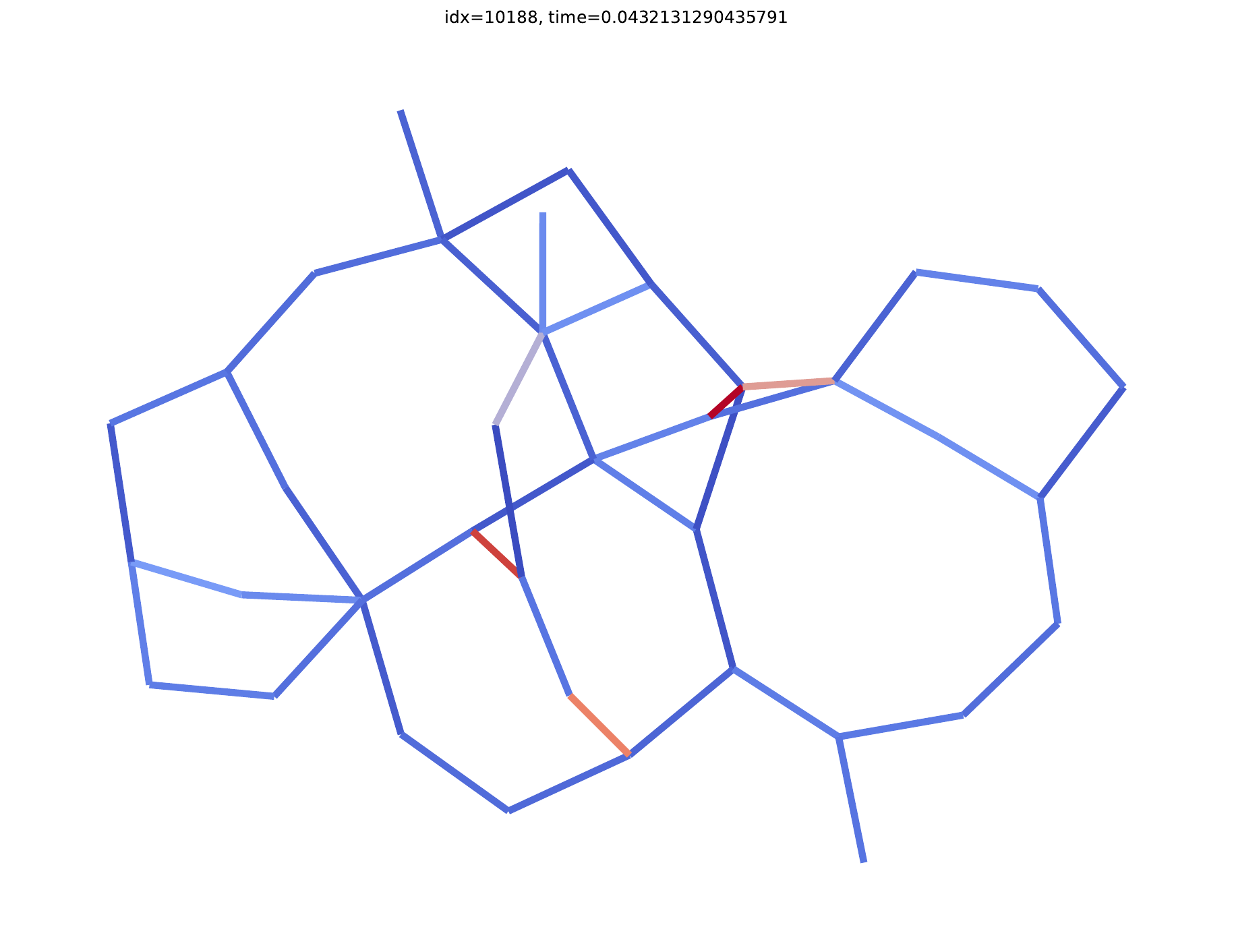} &
\imgcell{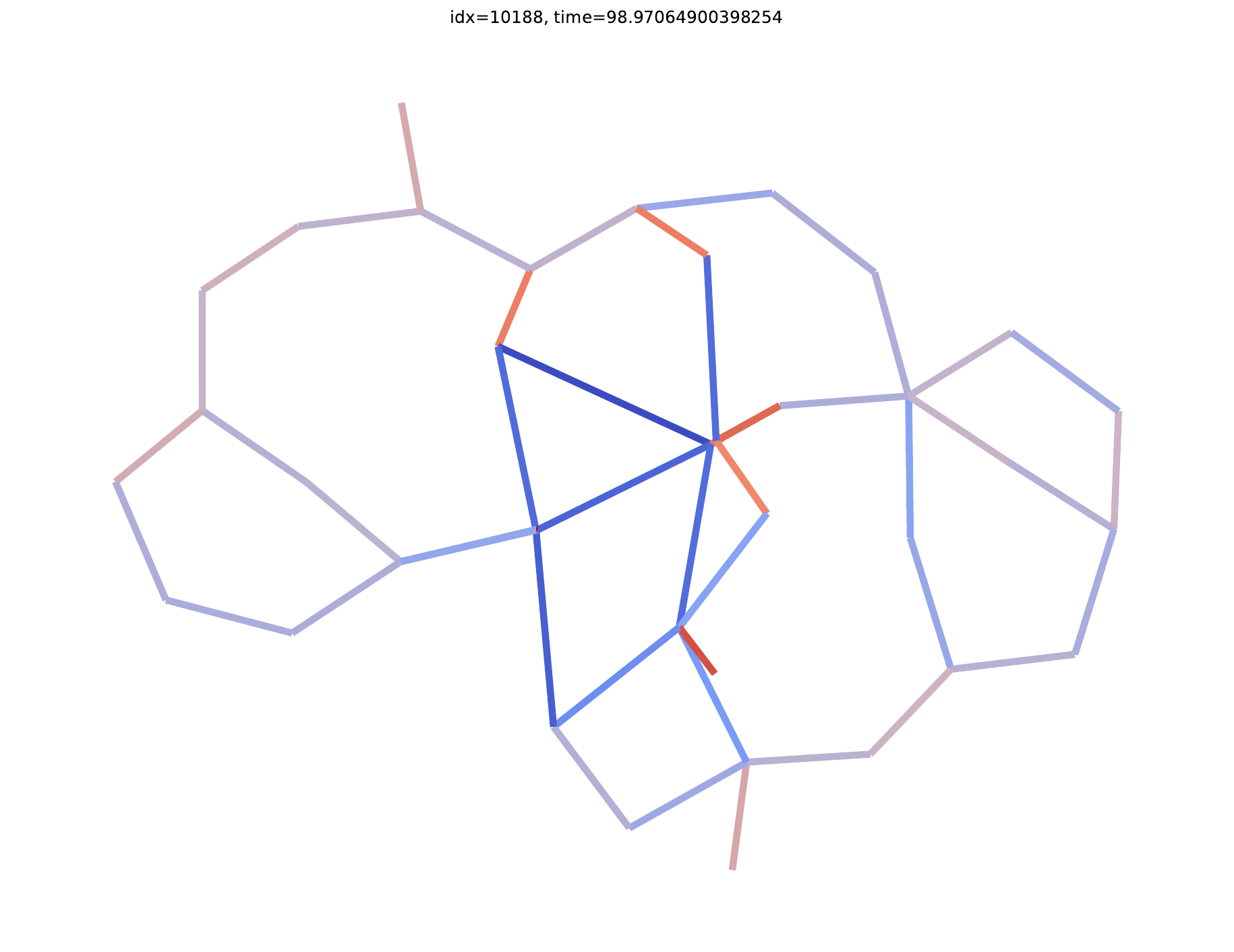} &
\imgcell{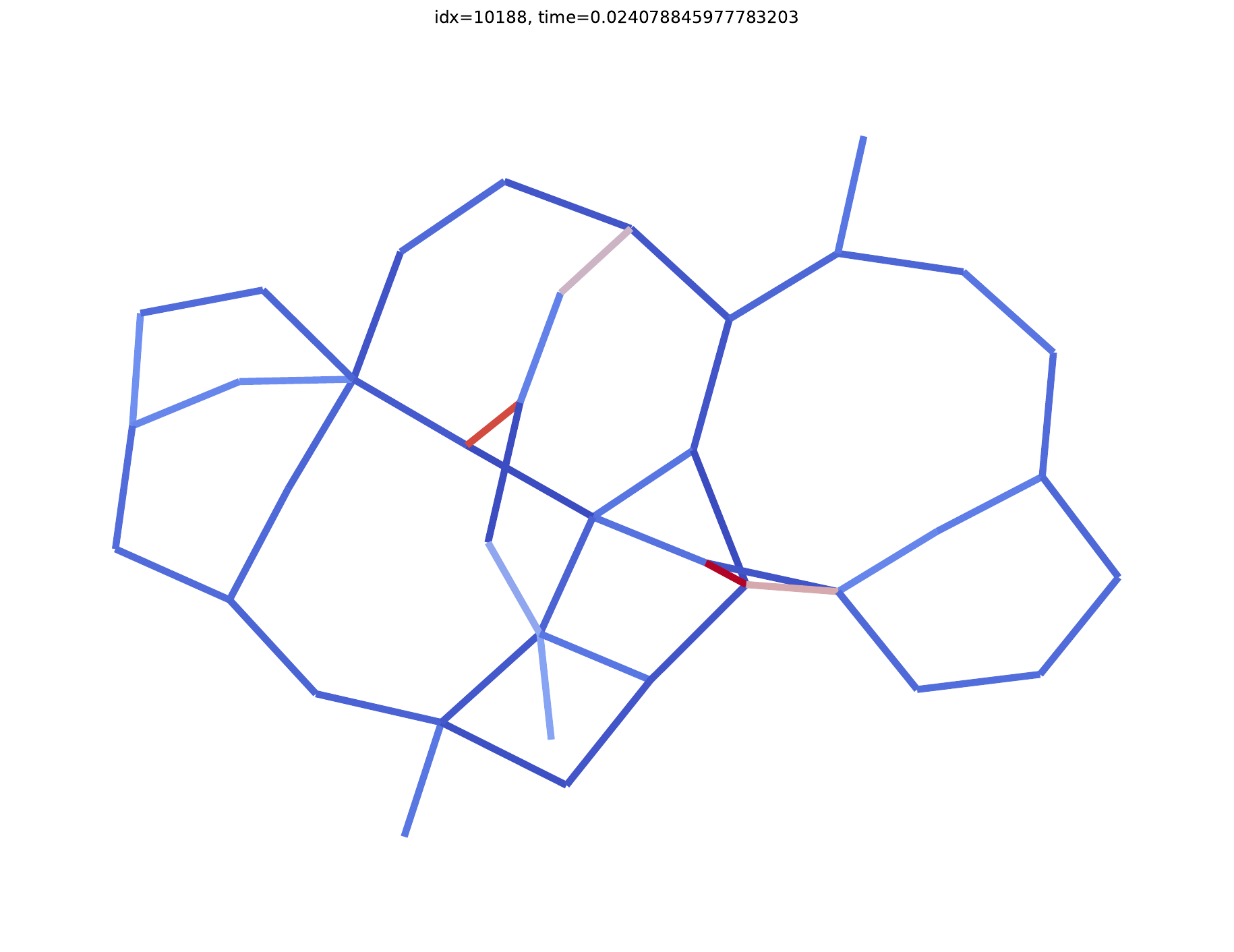} &
\imgcell{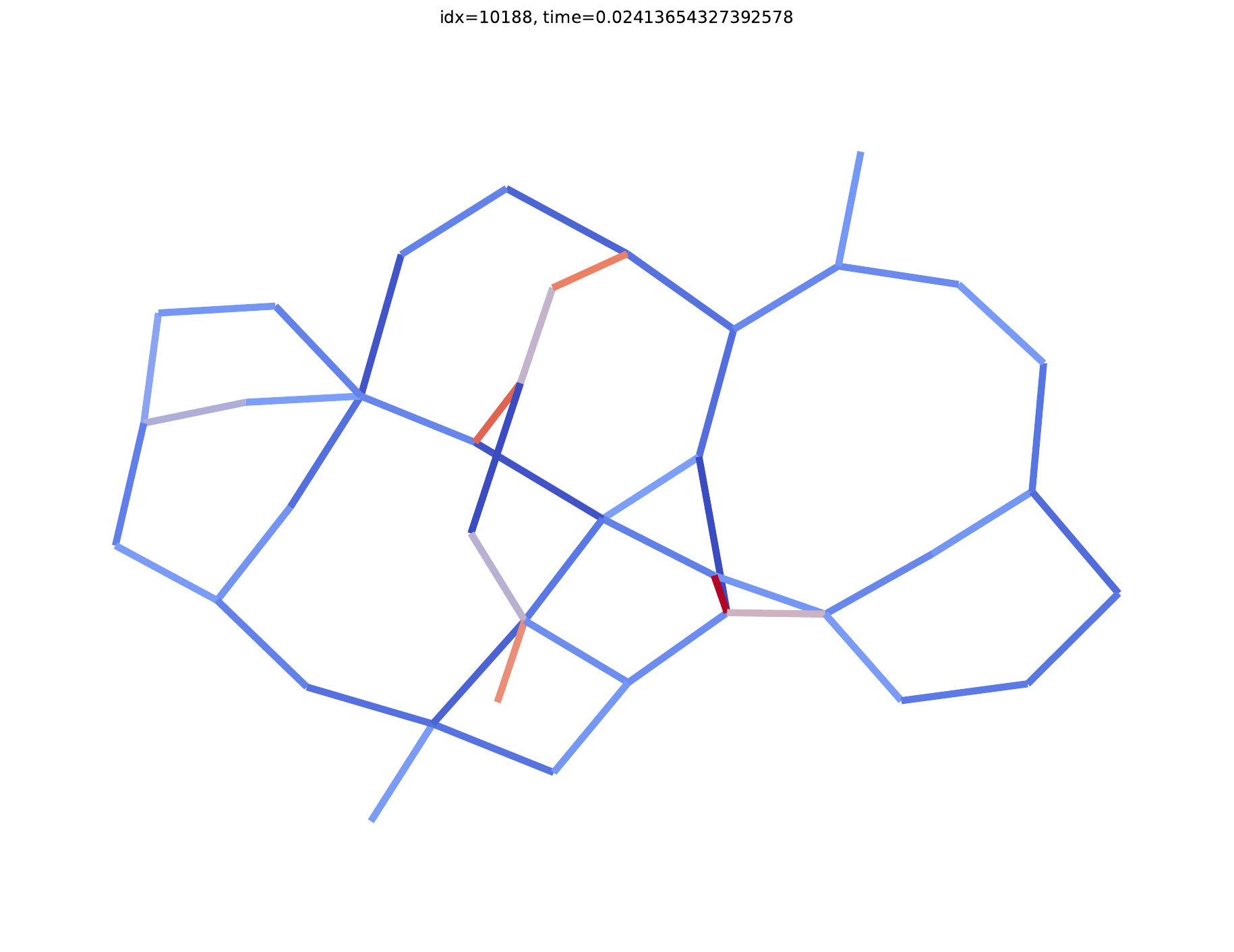} &
\imgcell{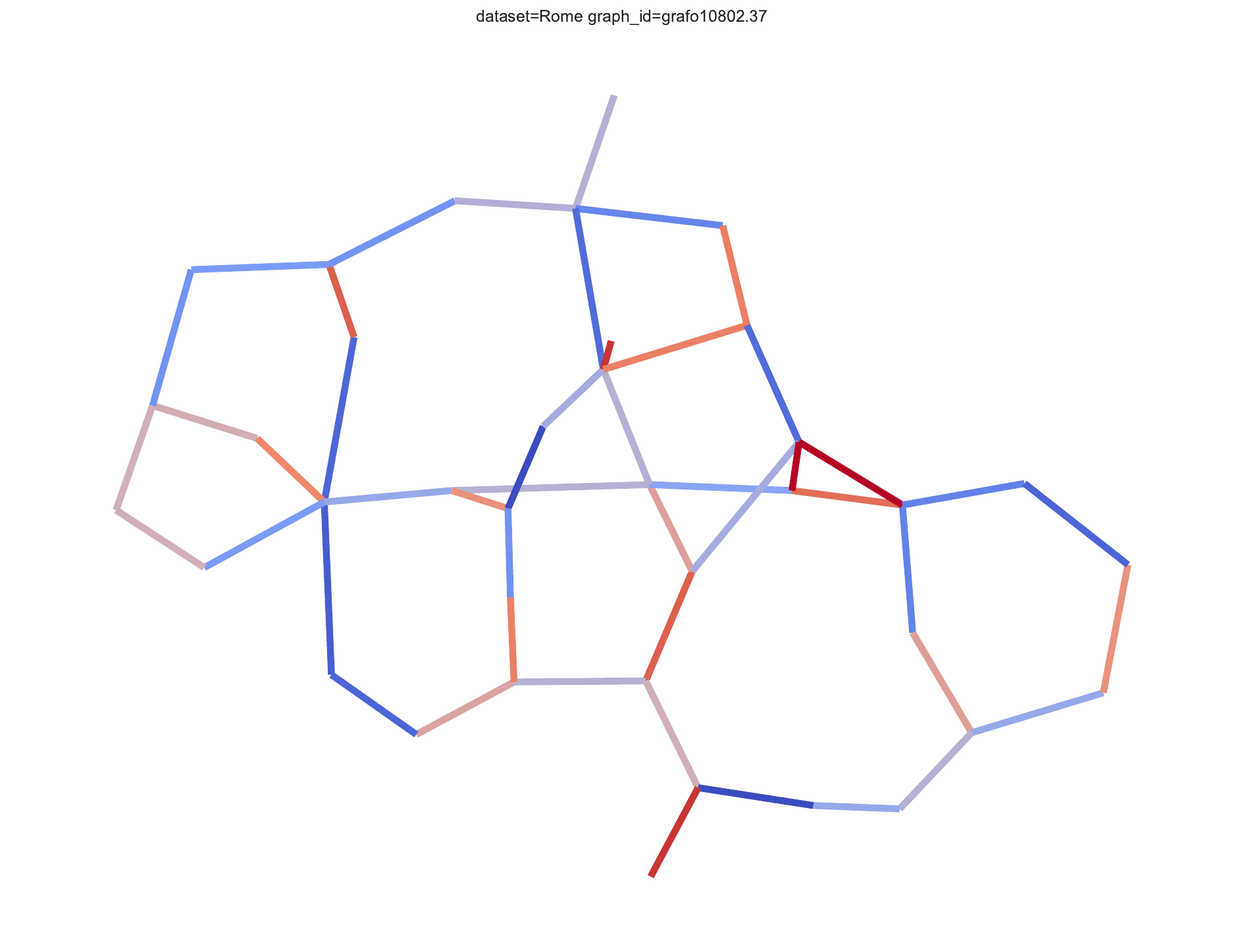} &
\imgcell{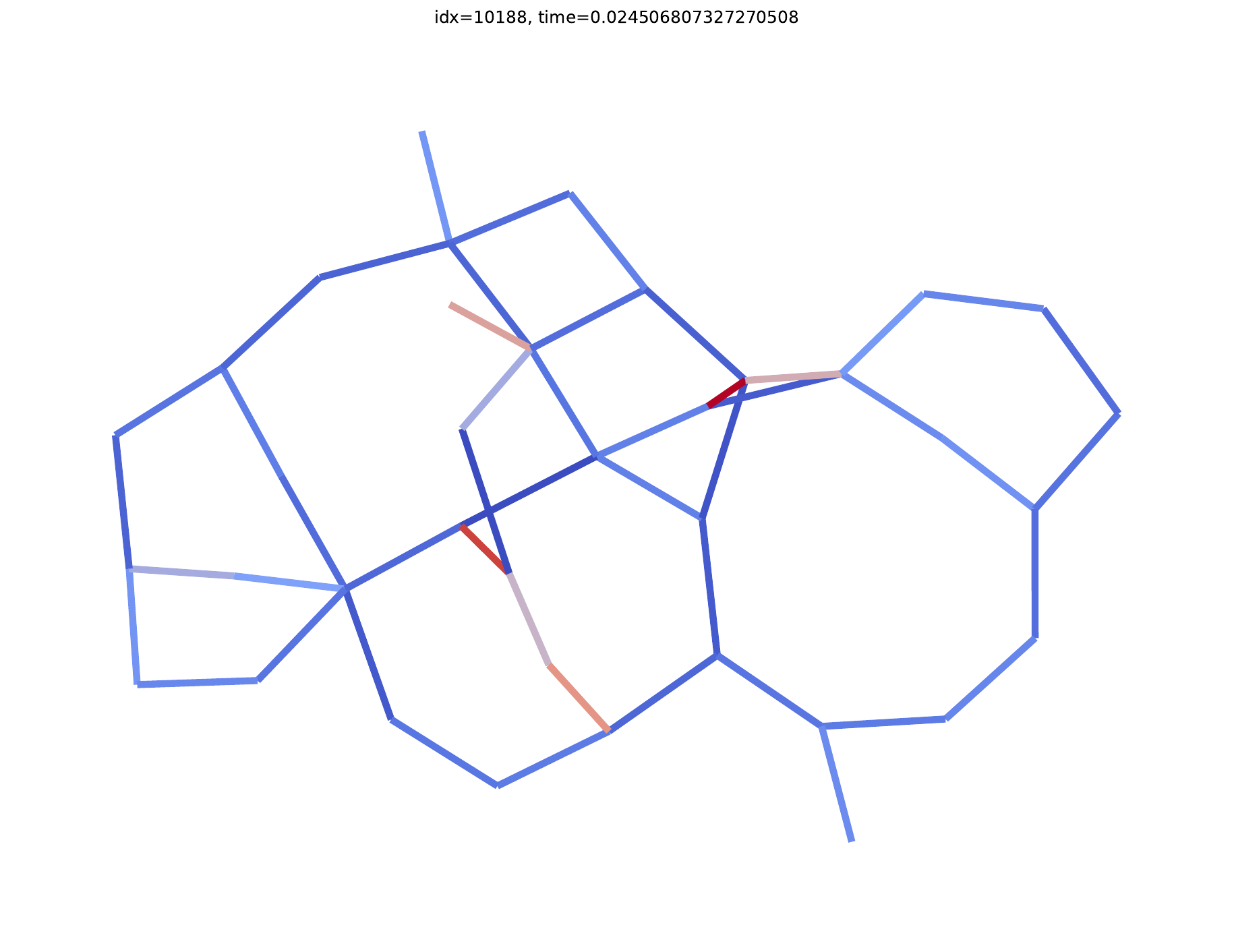} &
\imgcell{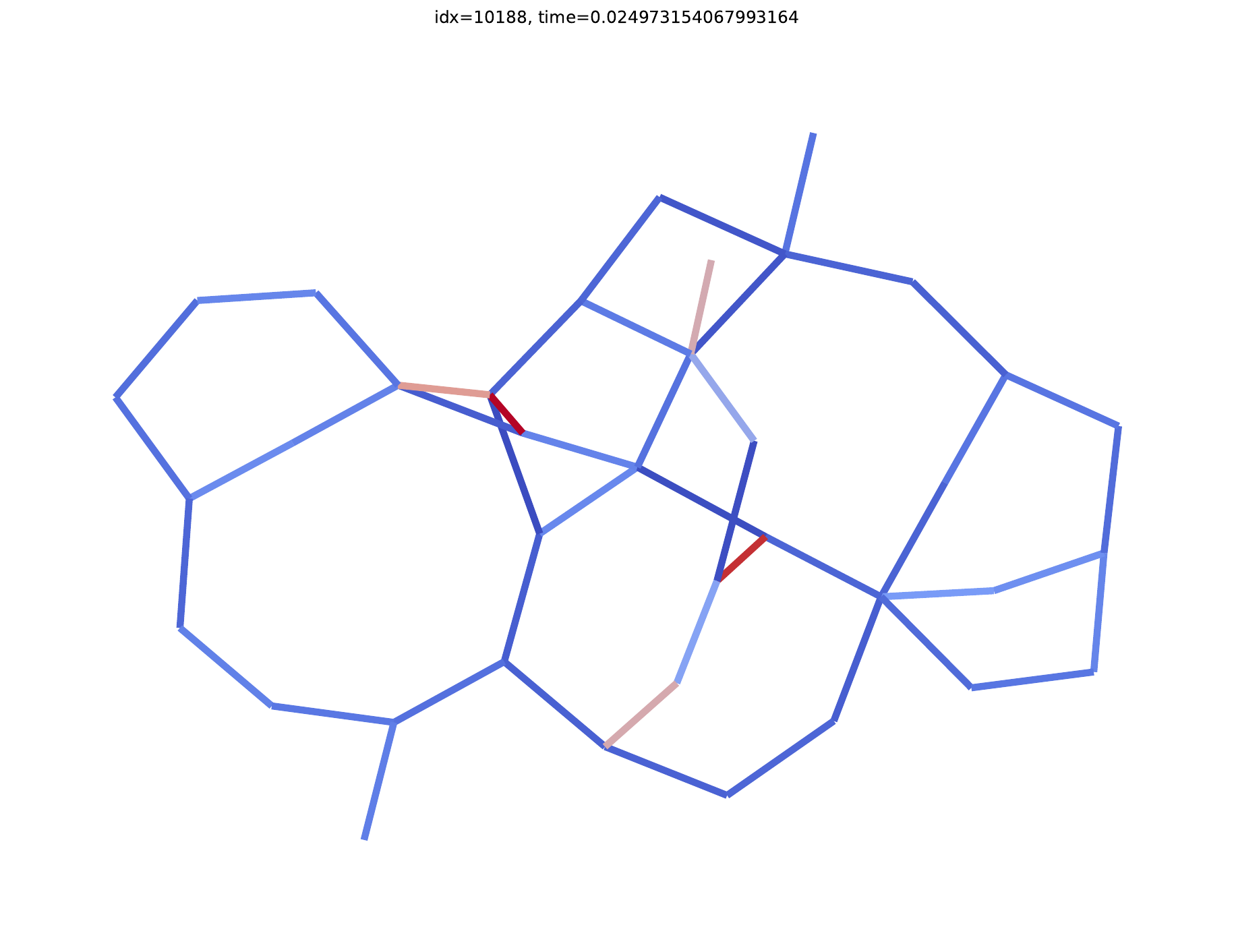} &
\imgcell{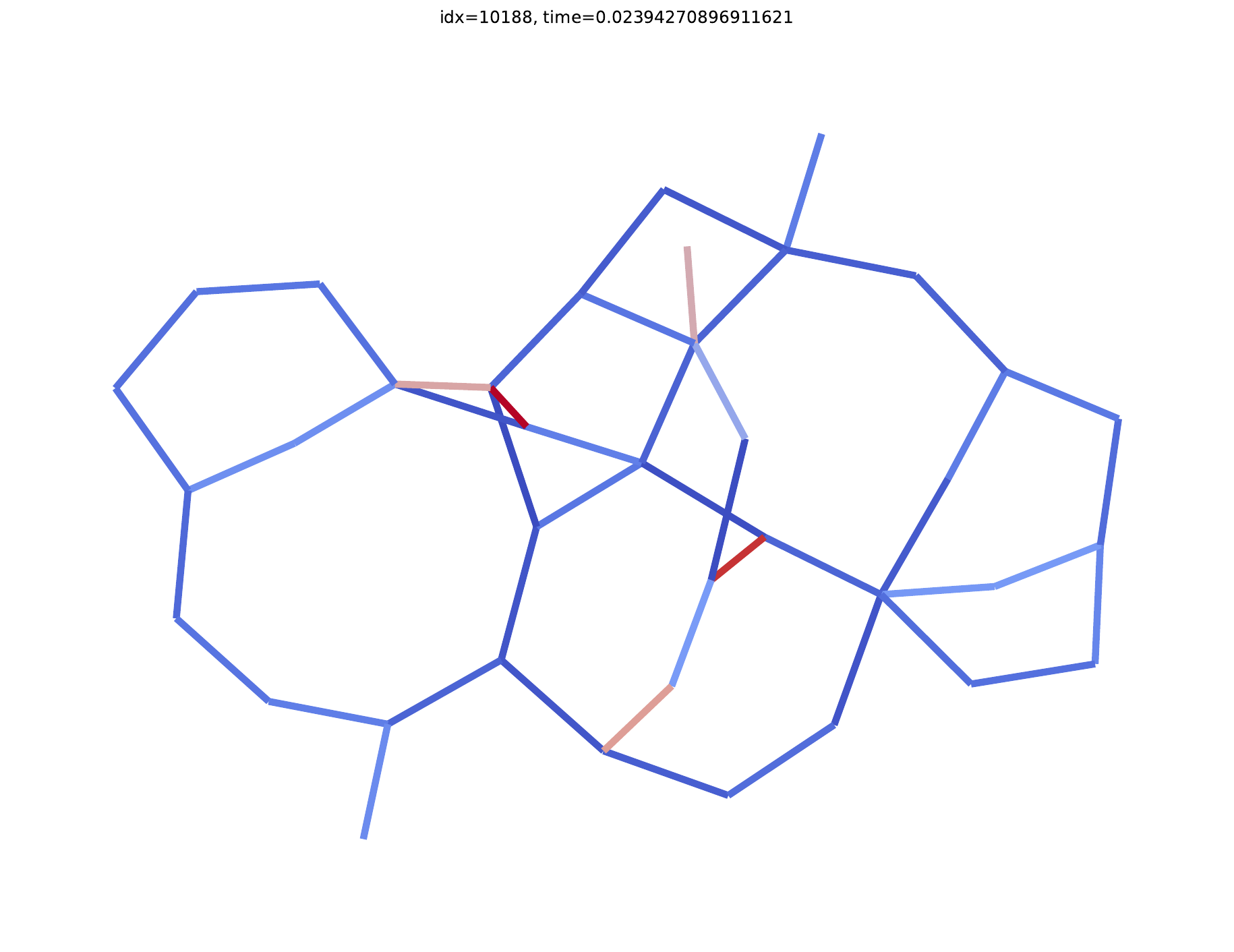} &
\imgcell{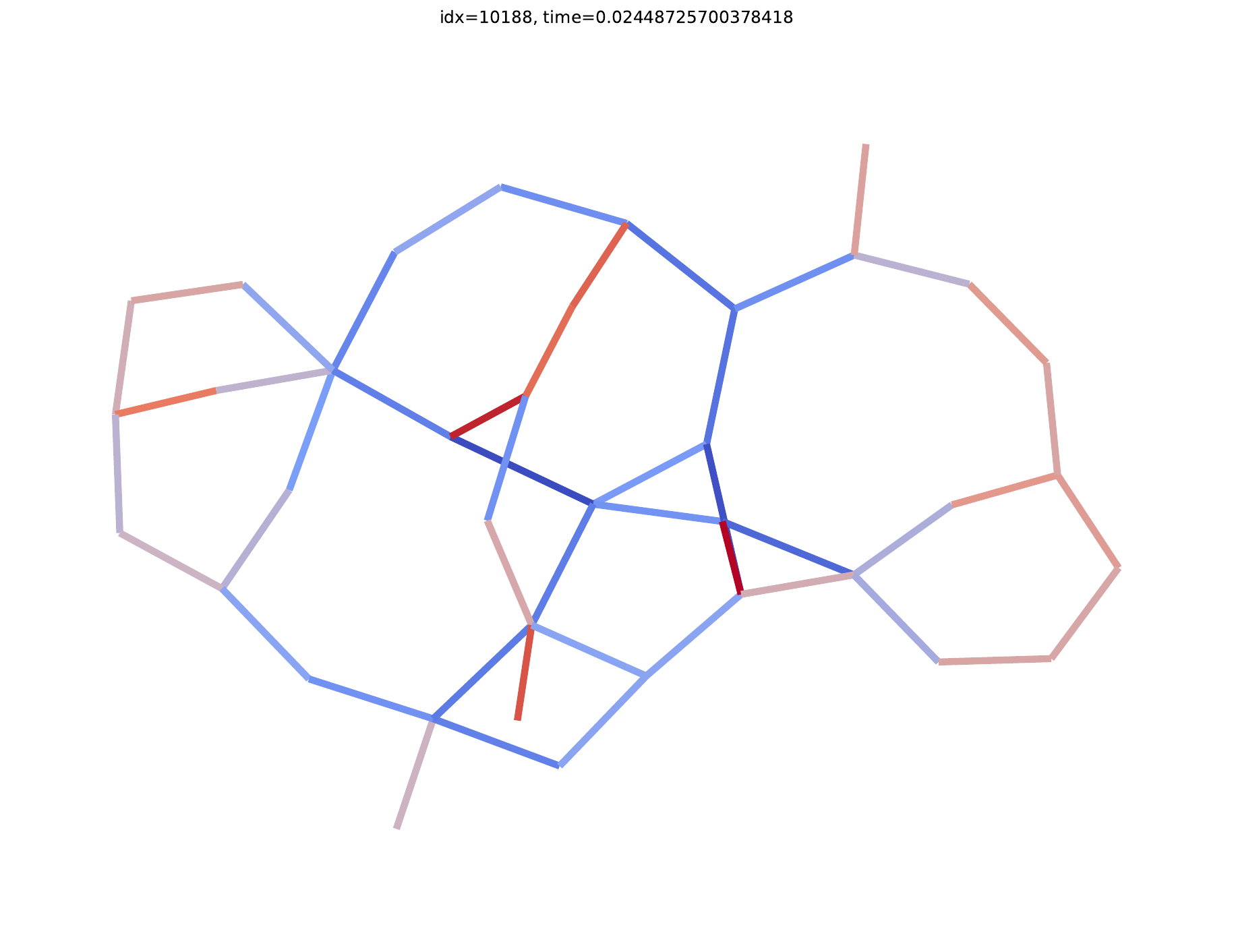} \\

&
t = 0.00s &
t = 0.38s &
t = 0.11s &
t = 0.04s &
t = 98.97s &
t = 0.02s &
t = 0.02s &
t = 0.02s &
t = 0.02s &
t = 0.02s &
t = 0.02s &
t = 0.02s \\

\makecell{\bfseries grafo7773.82\\N = 80\\M = 112} &
\imgcell{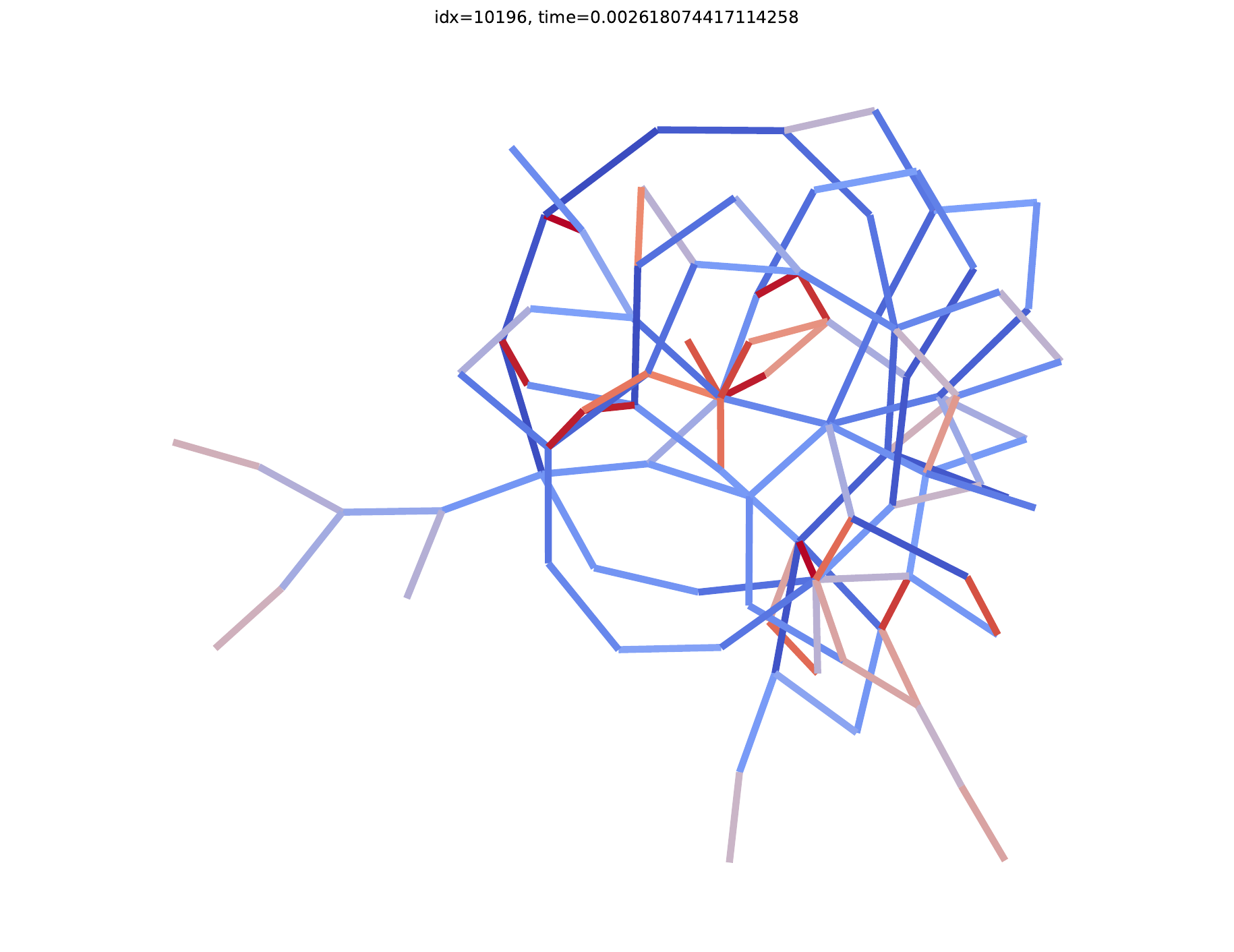} &
\imgcell{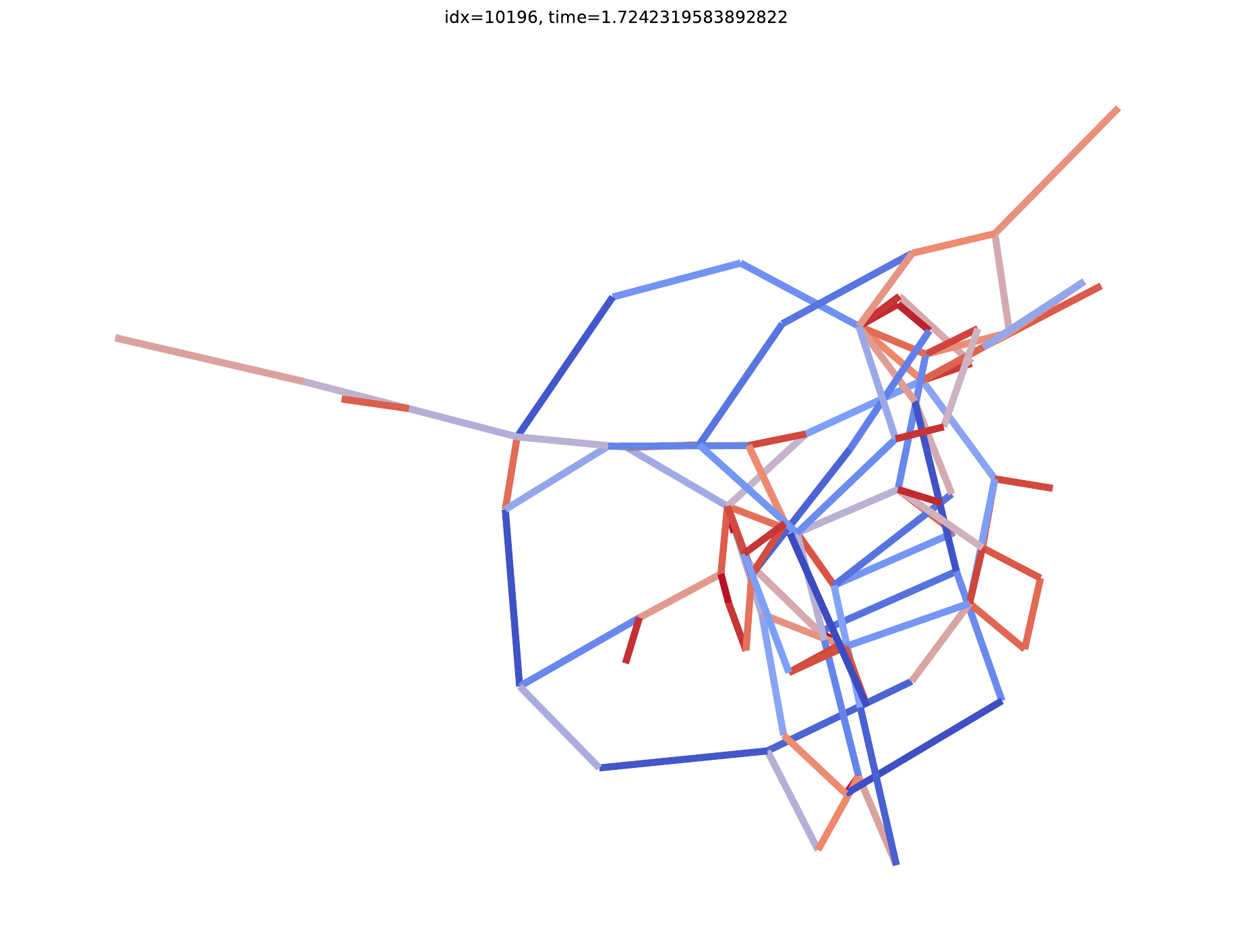} &
\imgcell{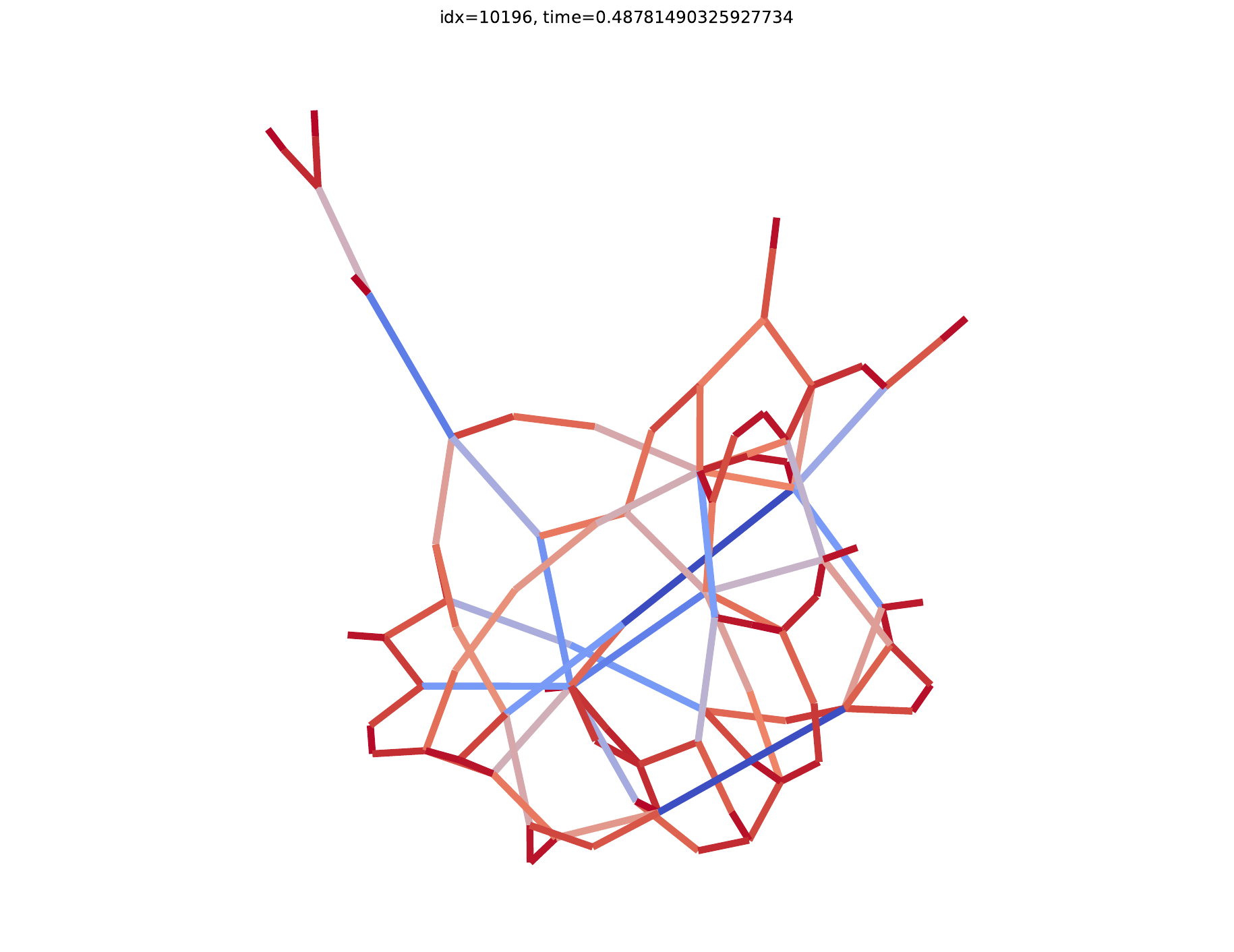} &
\imgcell{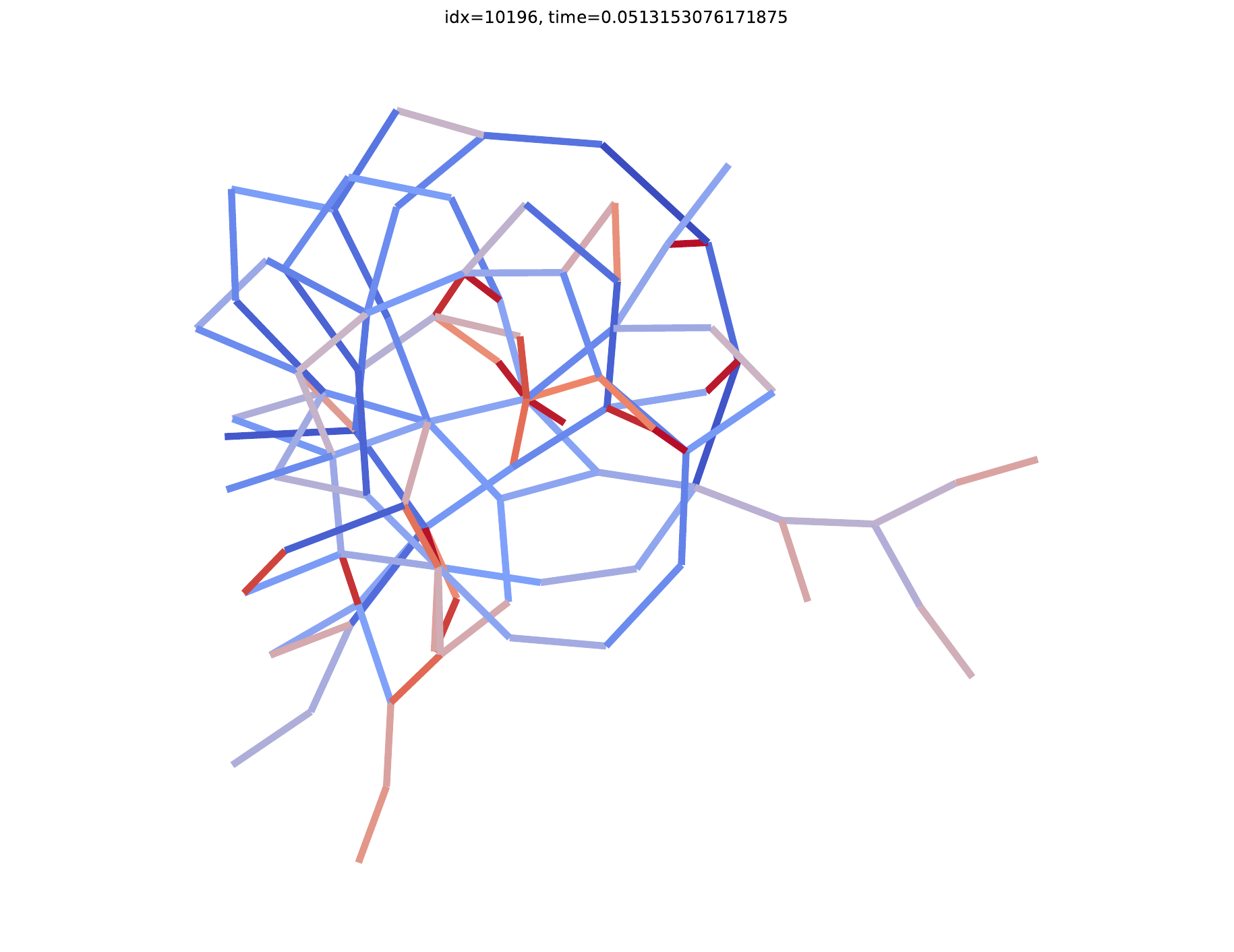} &
\imgcell{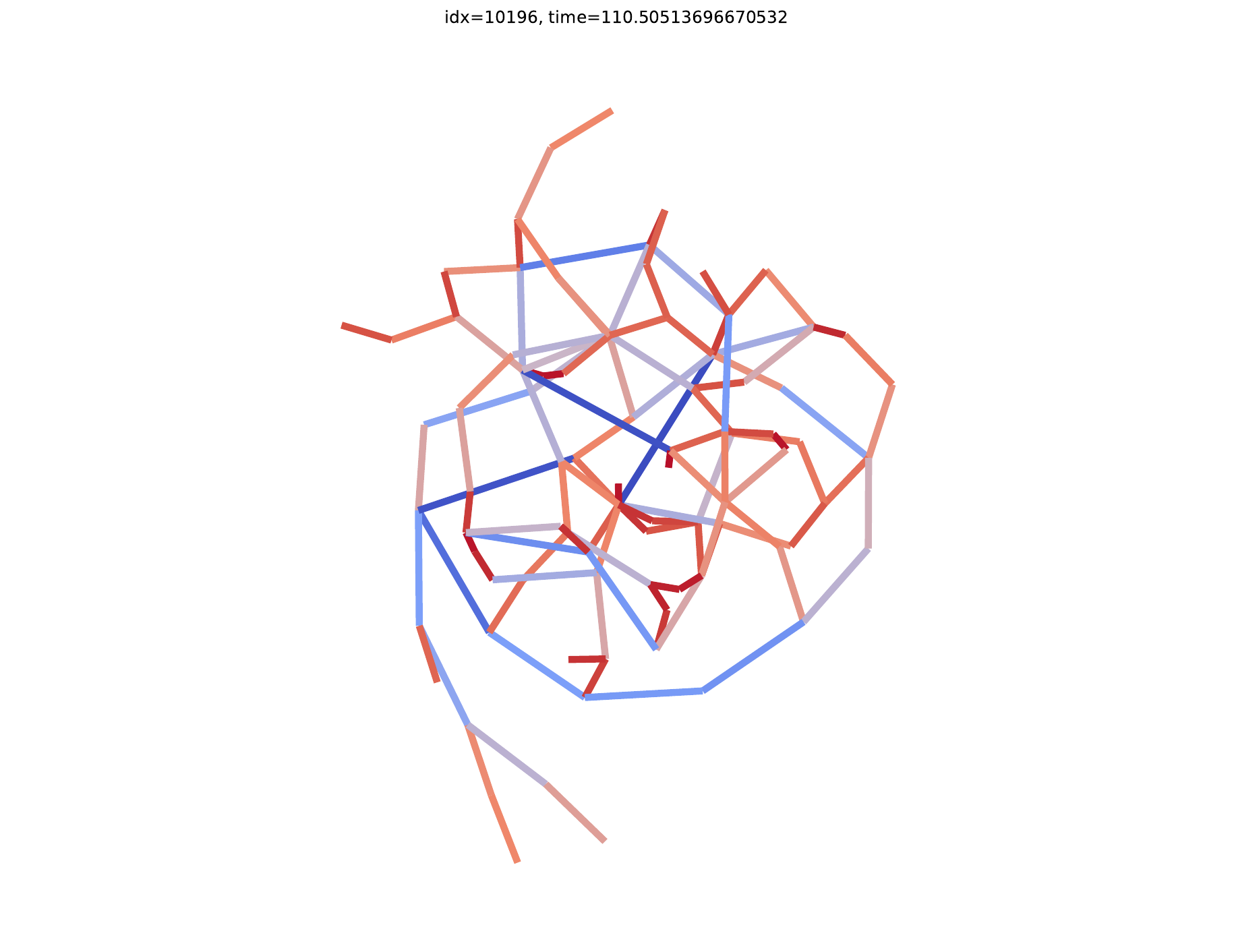} &
\imgcell{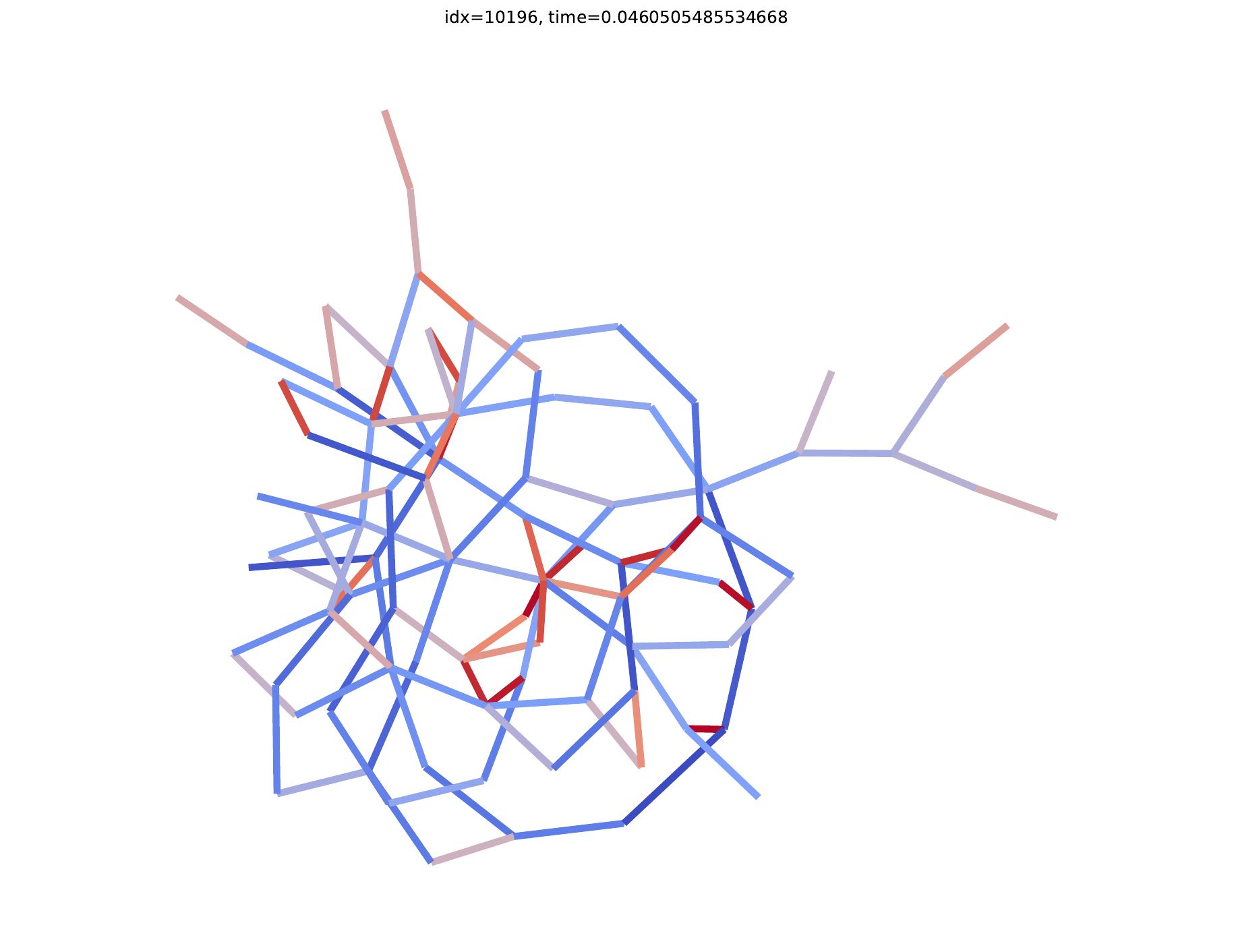} &
\imgcell{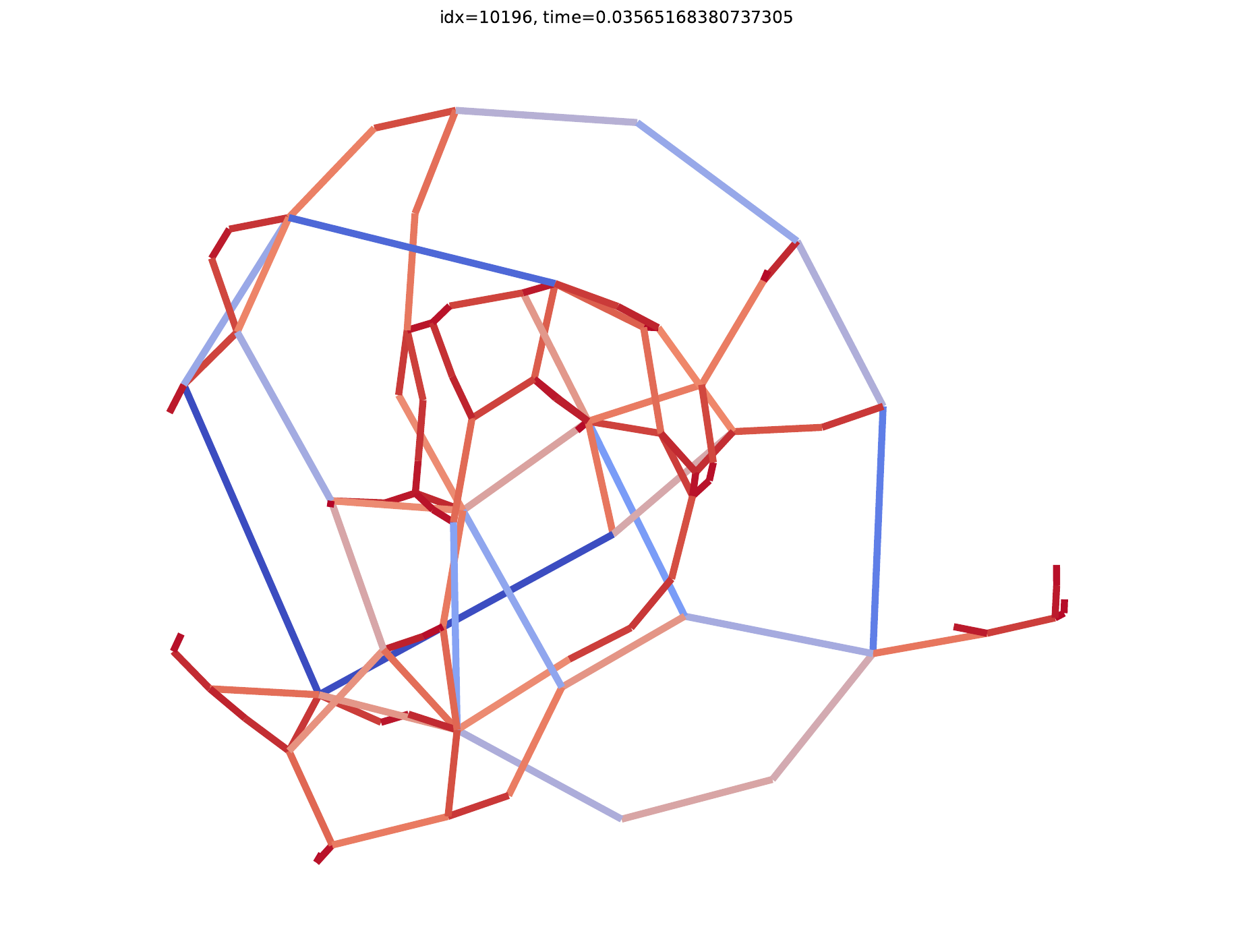} &
\imgcell{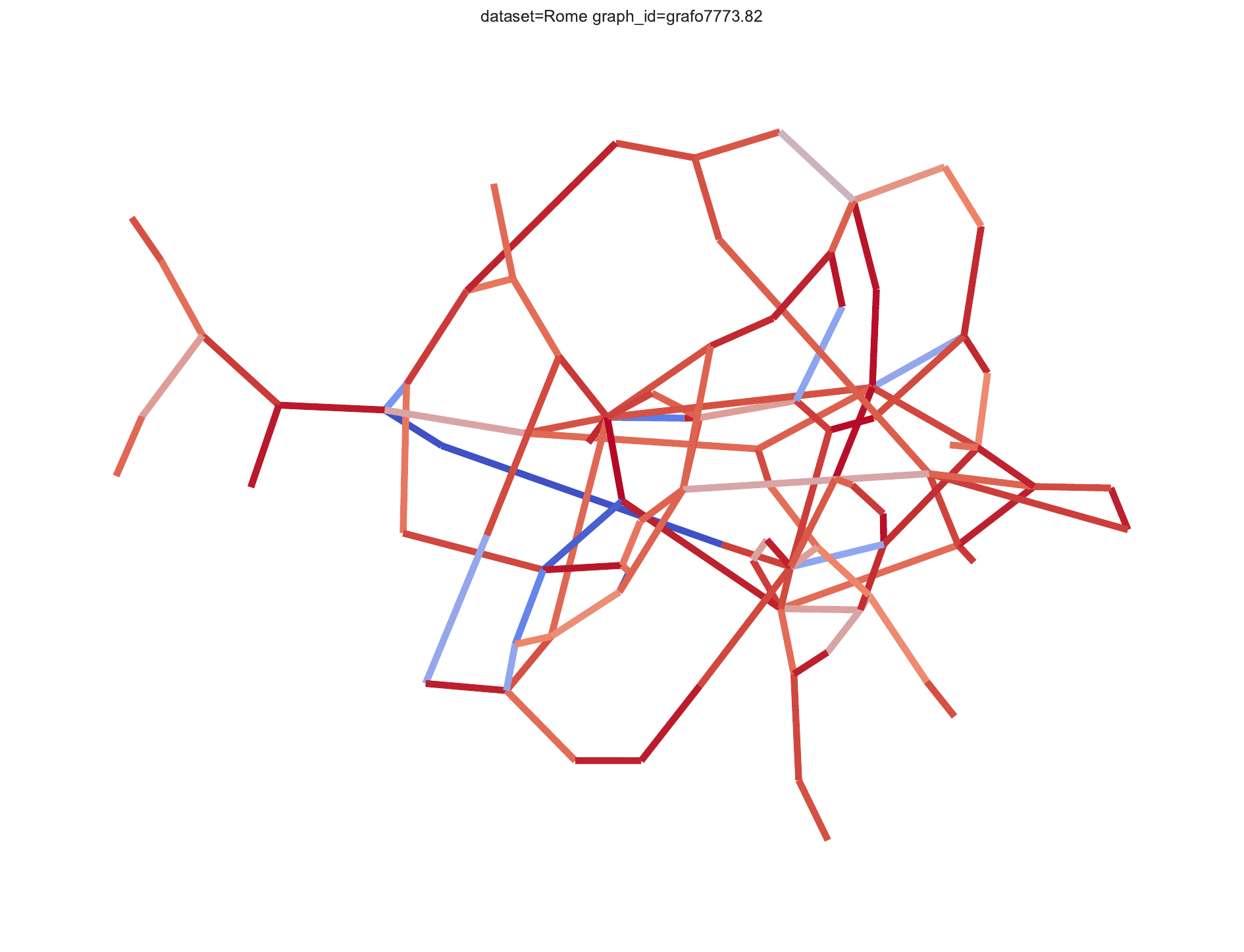} &
\imgcell{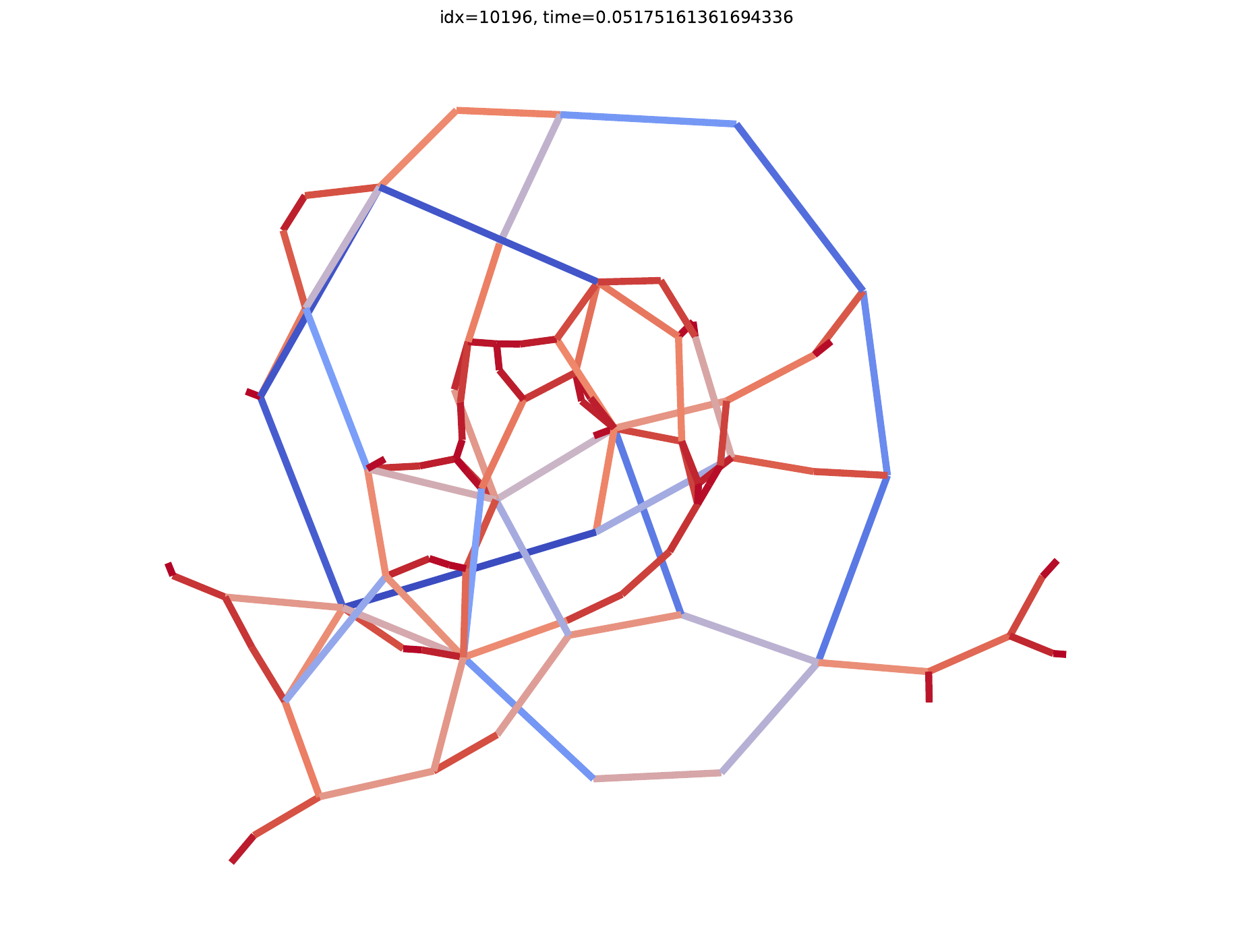} &
\imgcell{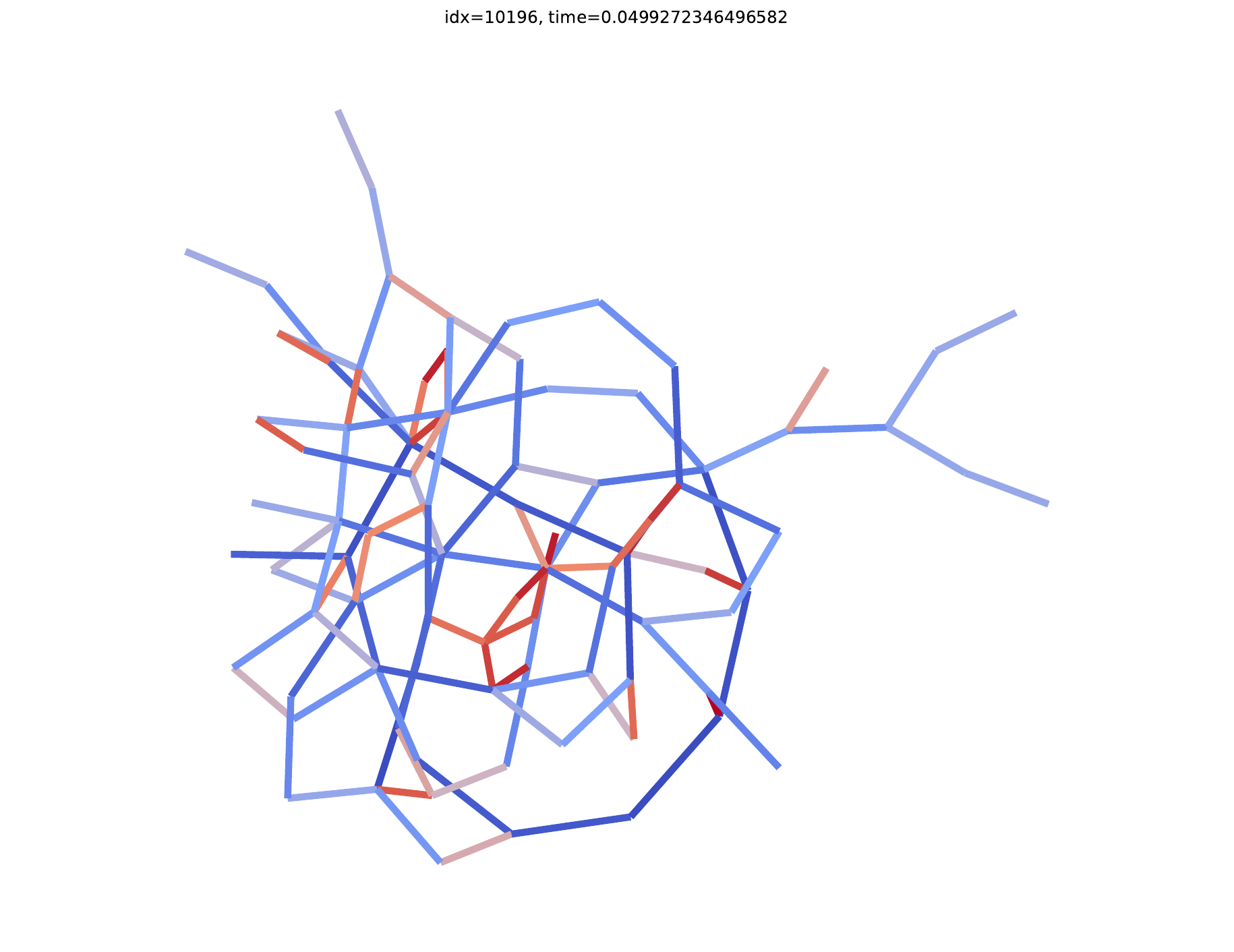} &
\imgcell{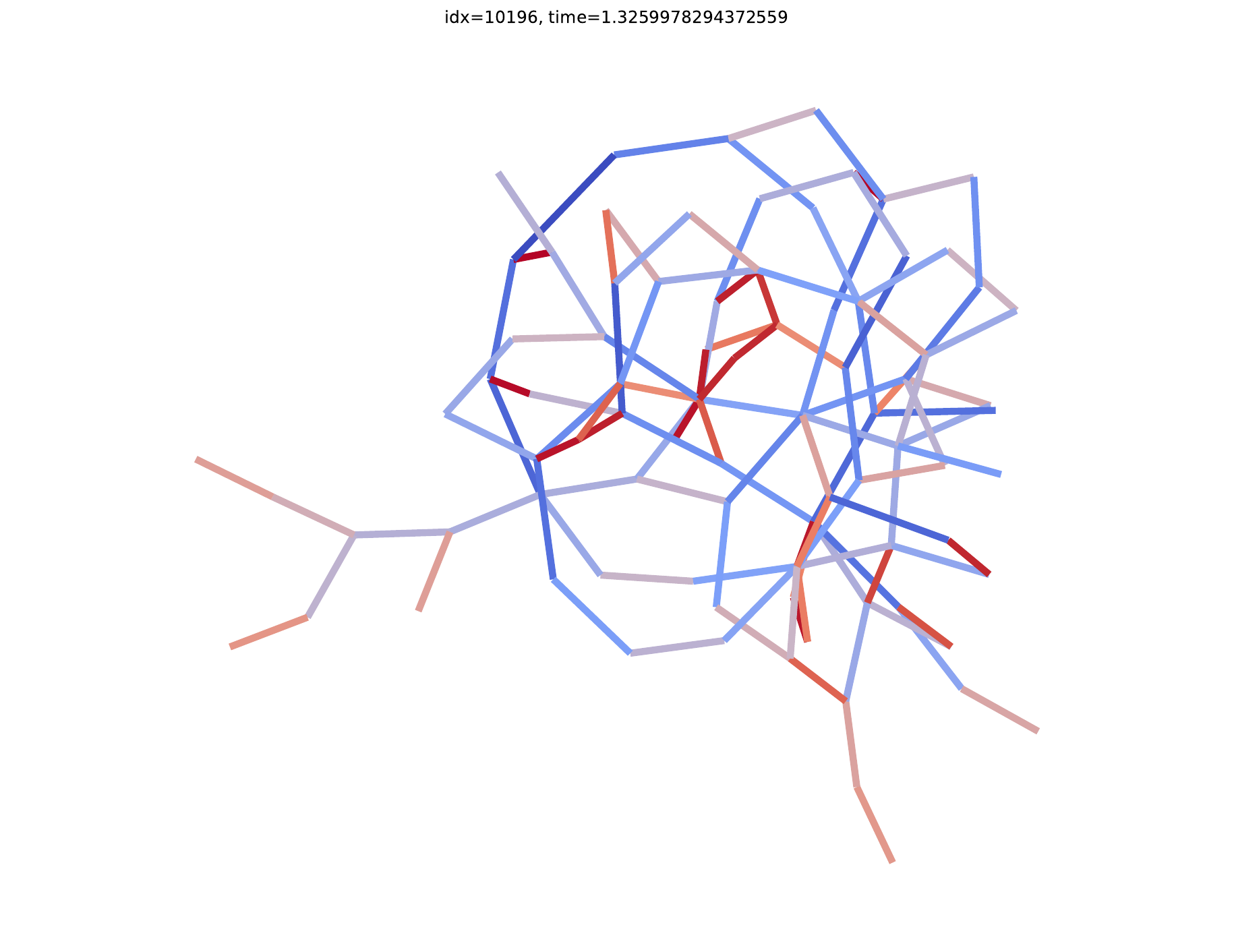} &
\imgcell{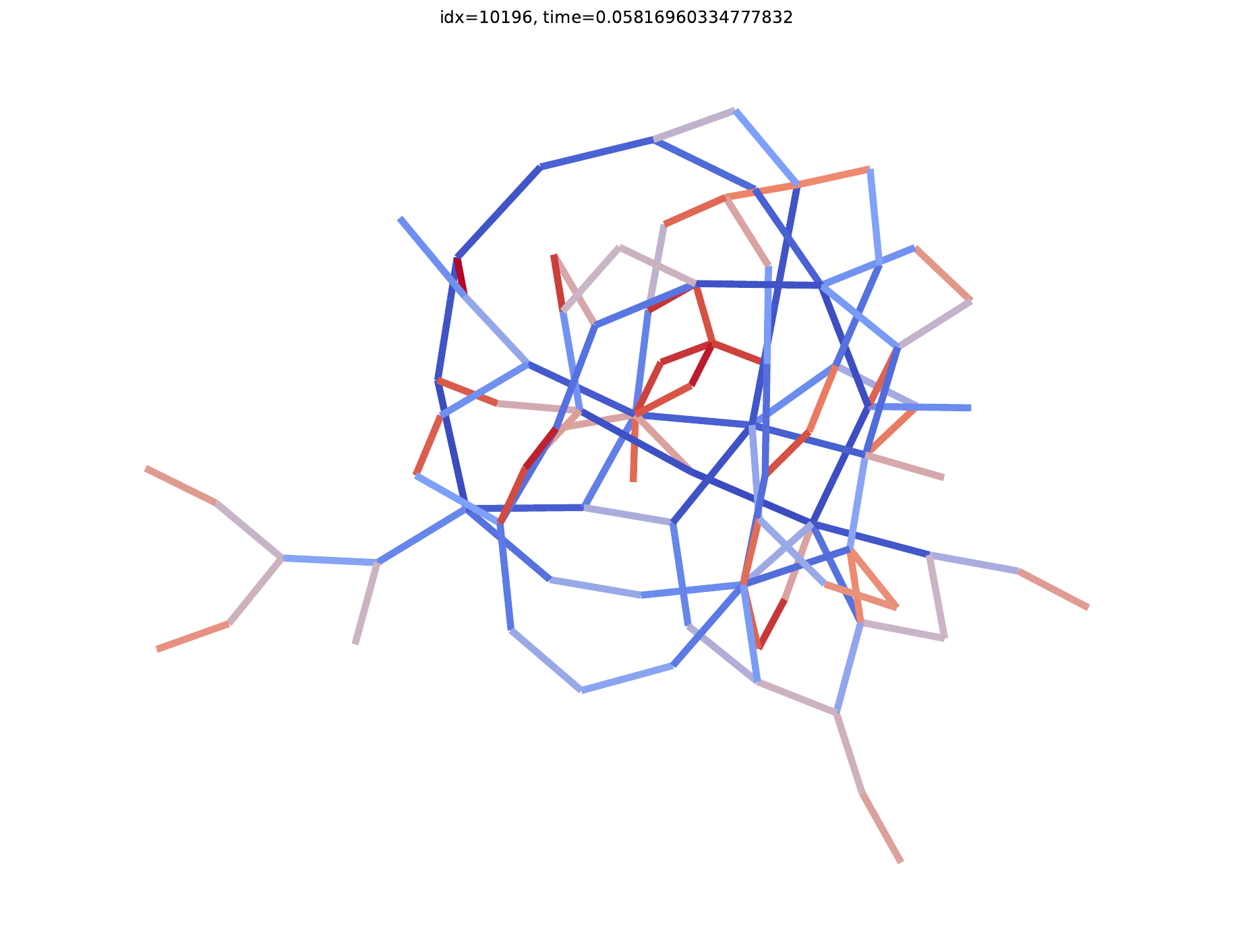} \\

&
t = 0.00s &
t = 1.72s &
t = 0.49s &
t = 0.05s &
t = 110.51s &
t = 0.05s &
t = 0.04s &
t = 0.05s &
t = 0.05s &
t = 0.05s &
t = 0.04s &
t = 0.06s \\

\makecell{\bfseries grafo1103.24\\N = 69\\M = 95} &
\imgcell{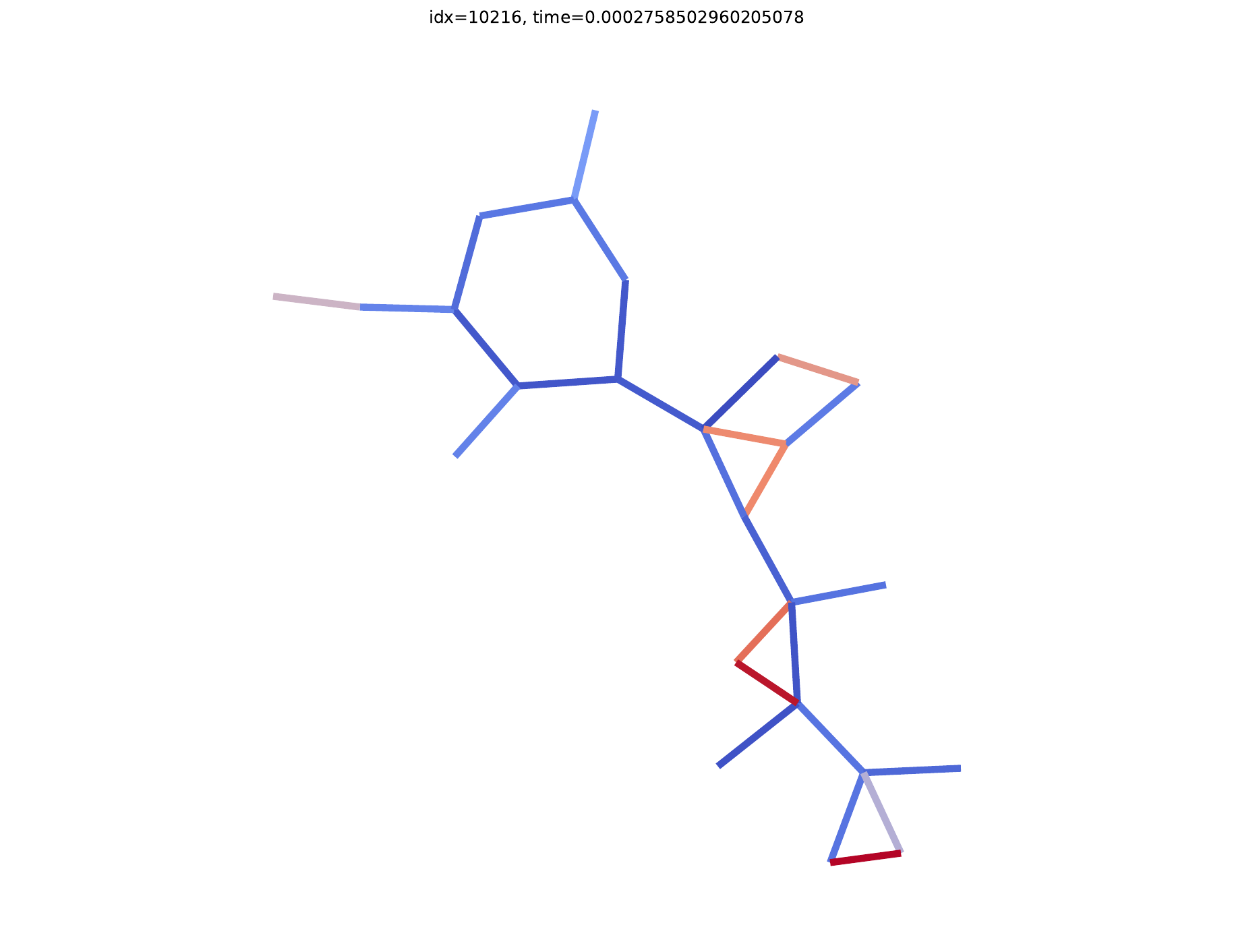} &
\imgcell{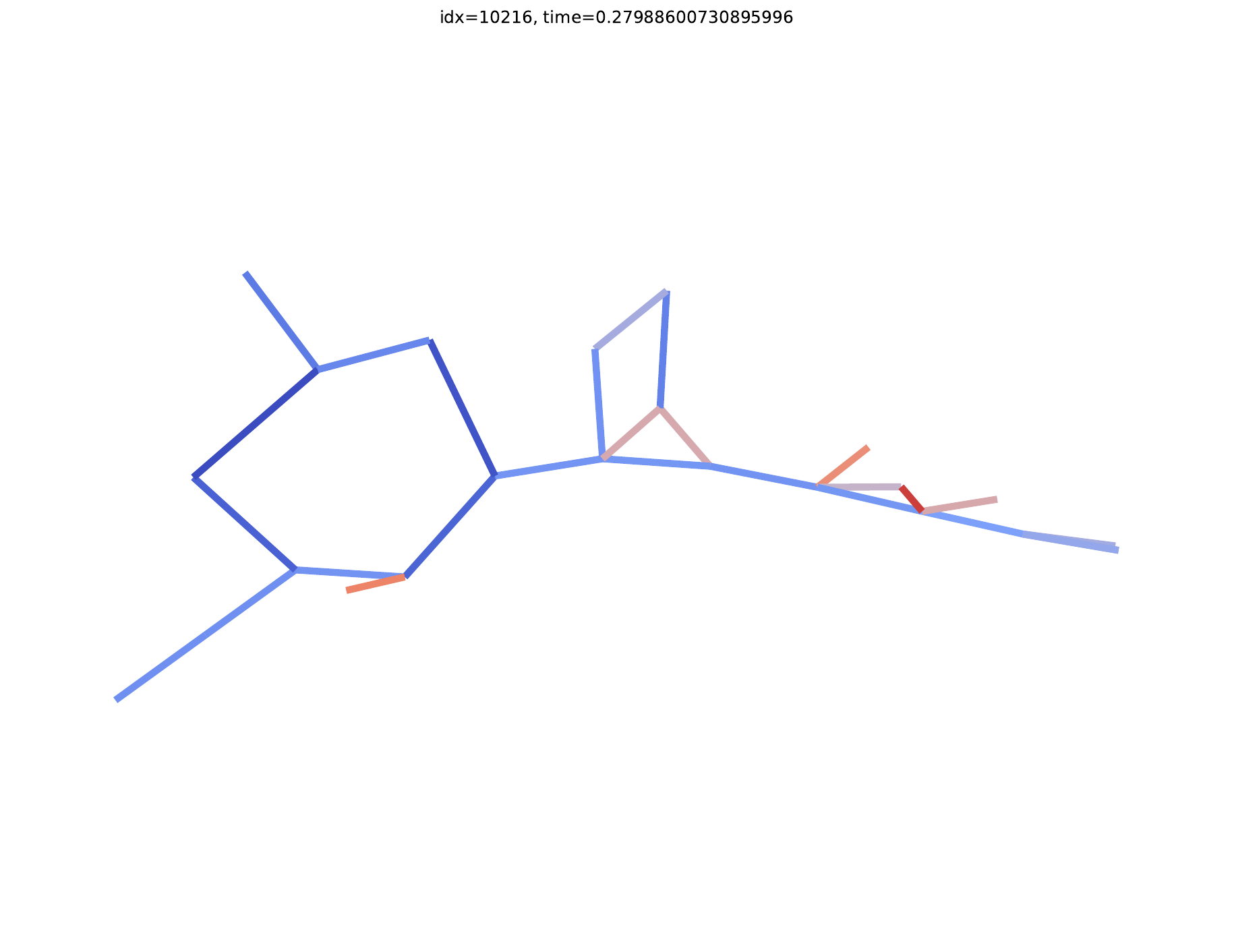} &
\imgcell{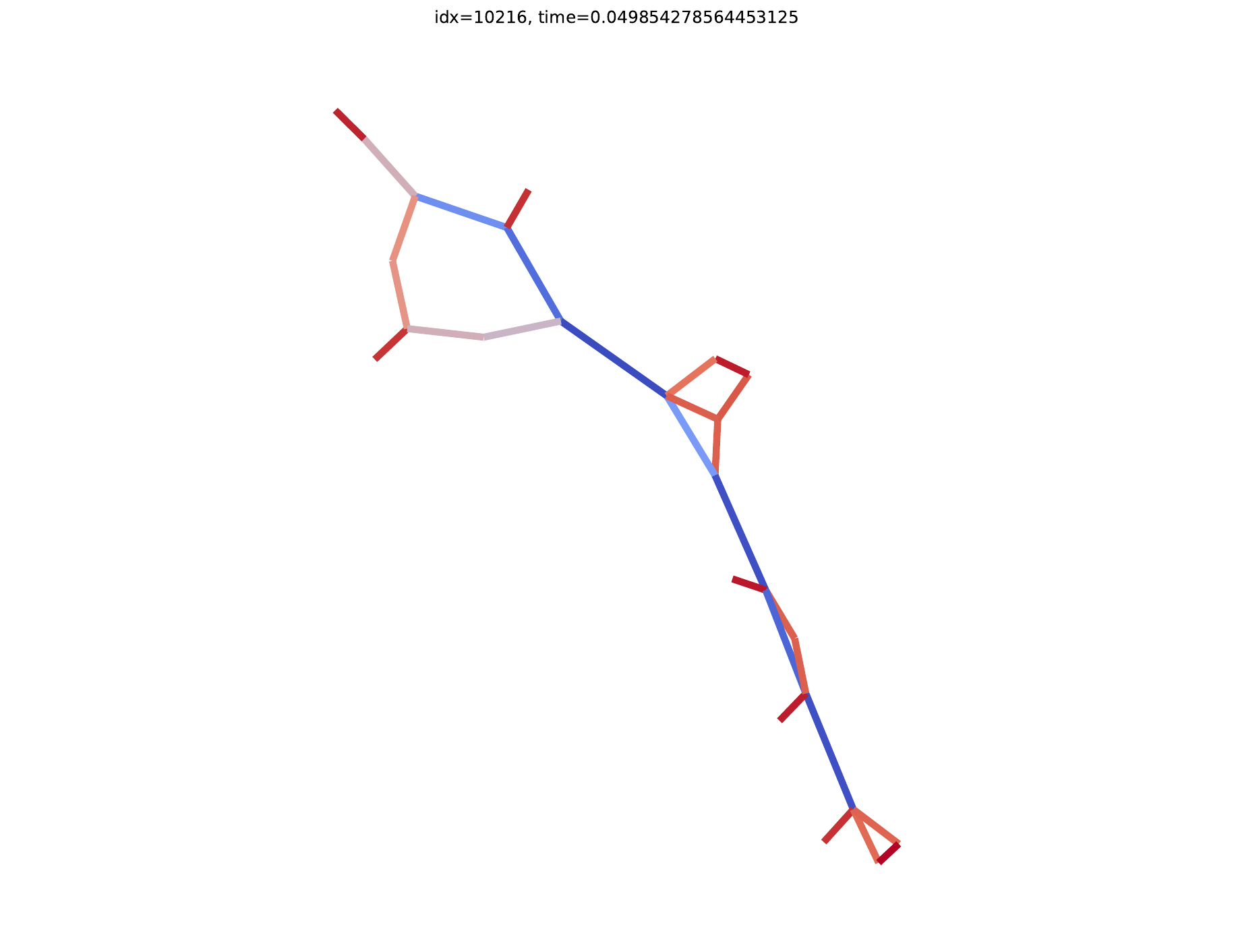} &
\imgcell{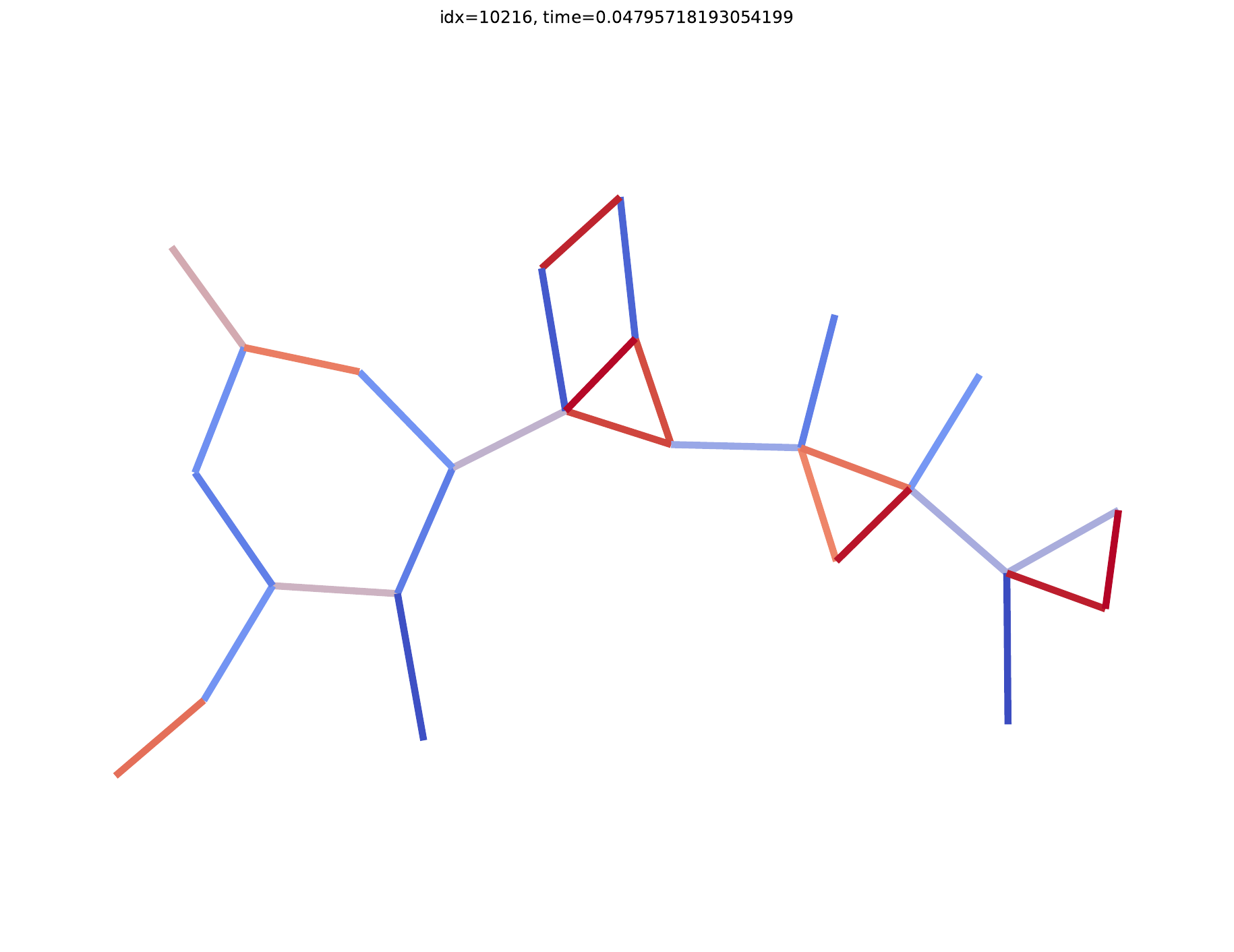} &
\imgcell{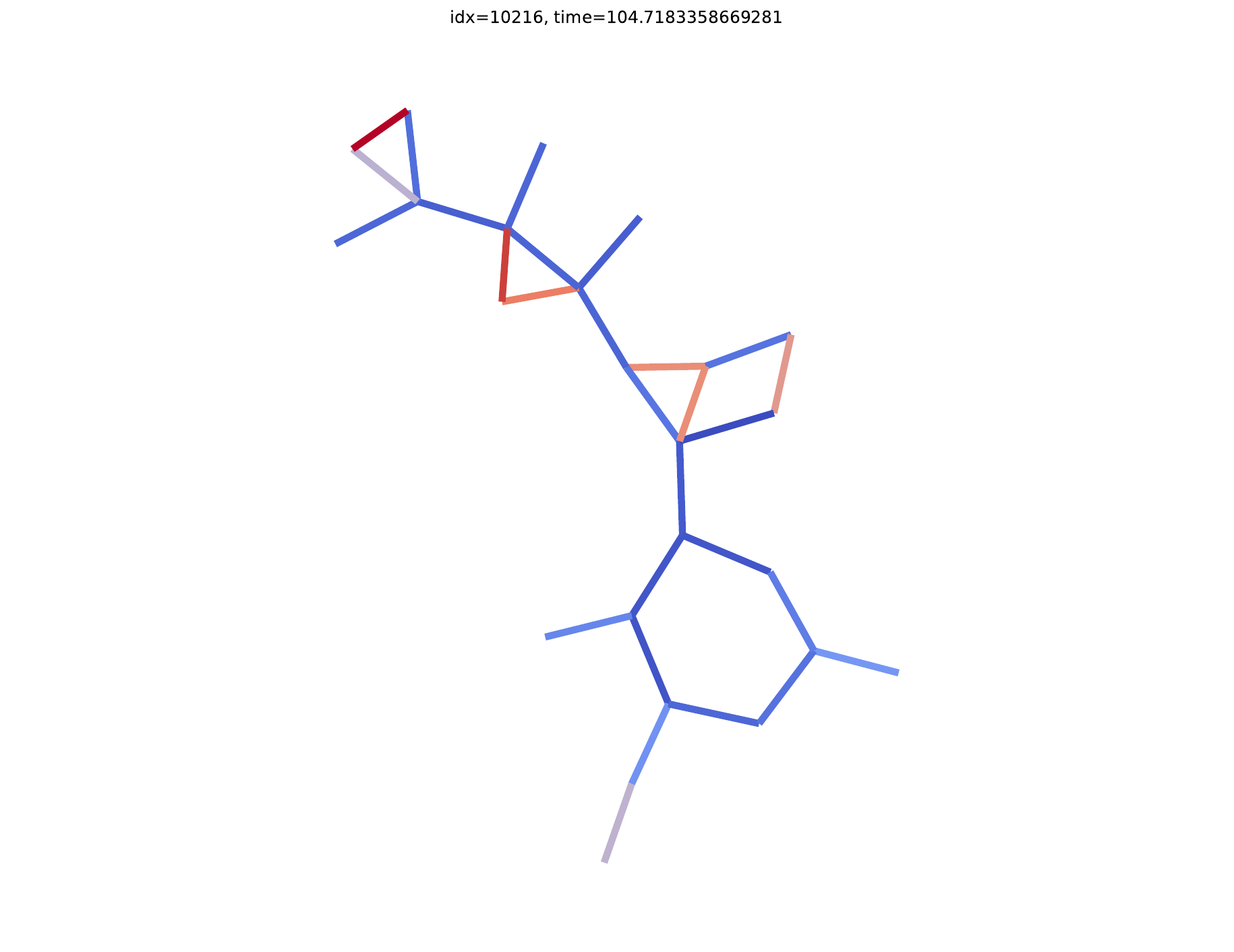} &
\imgcell{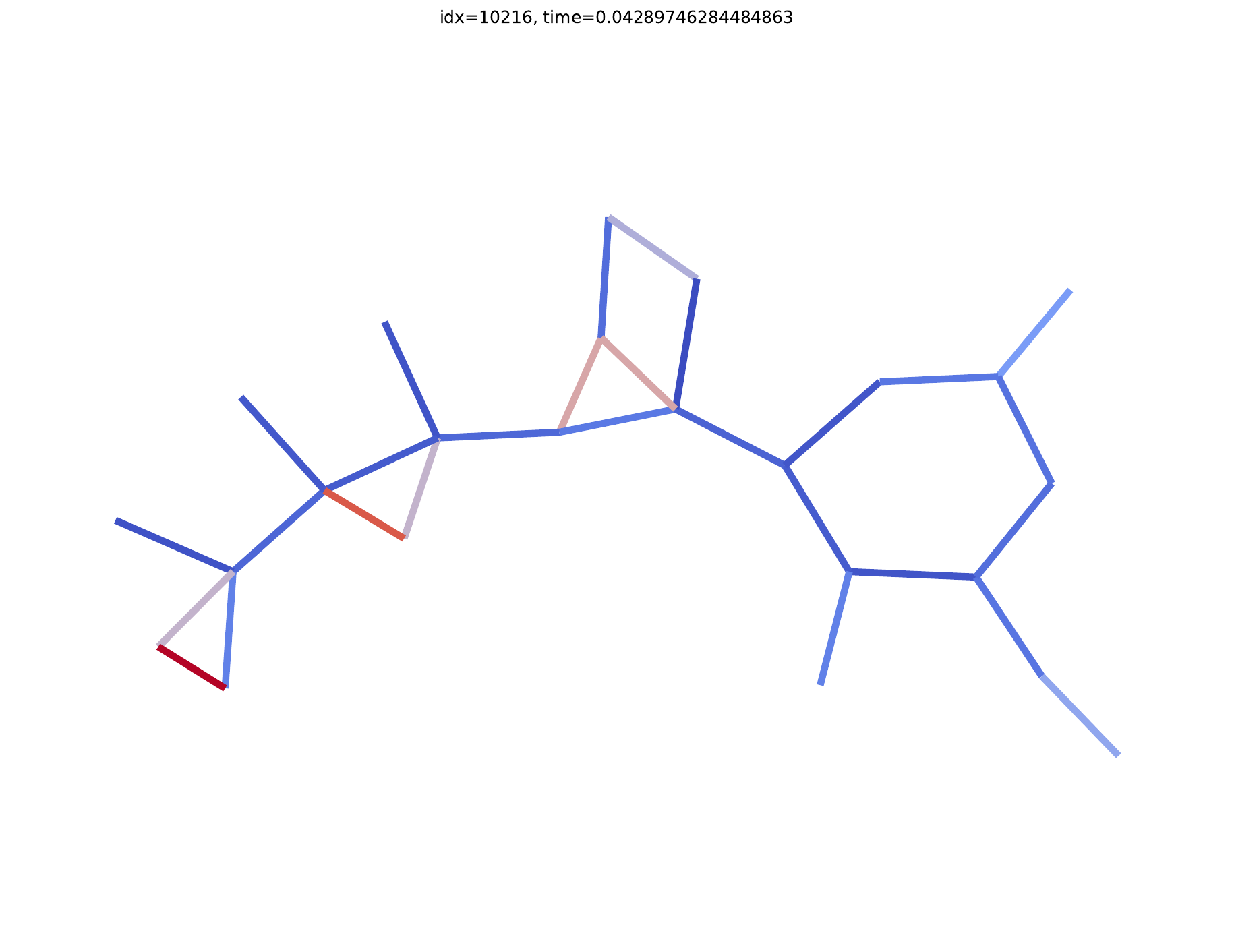} &
\imgcell{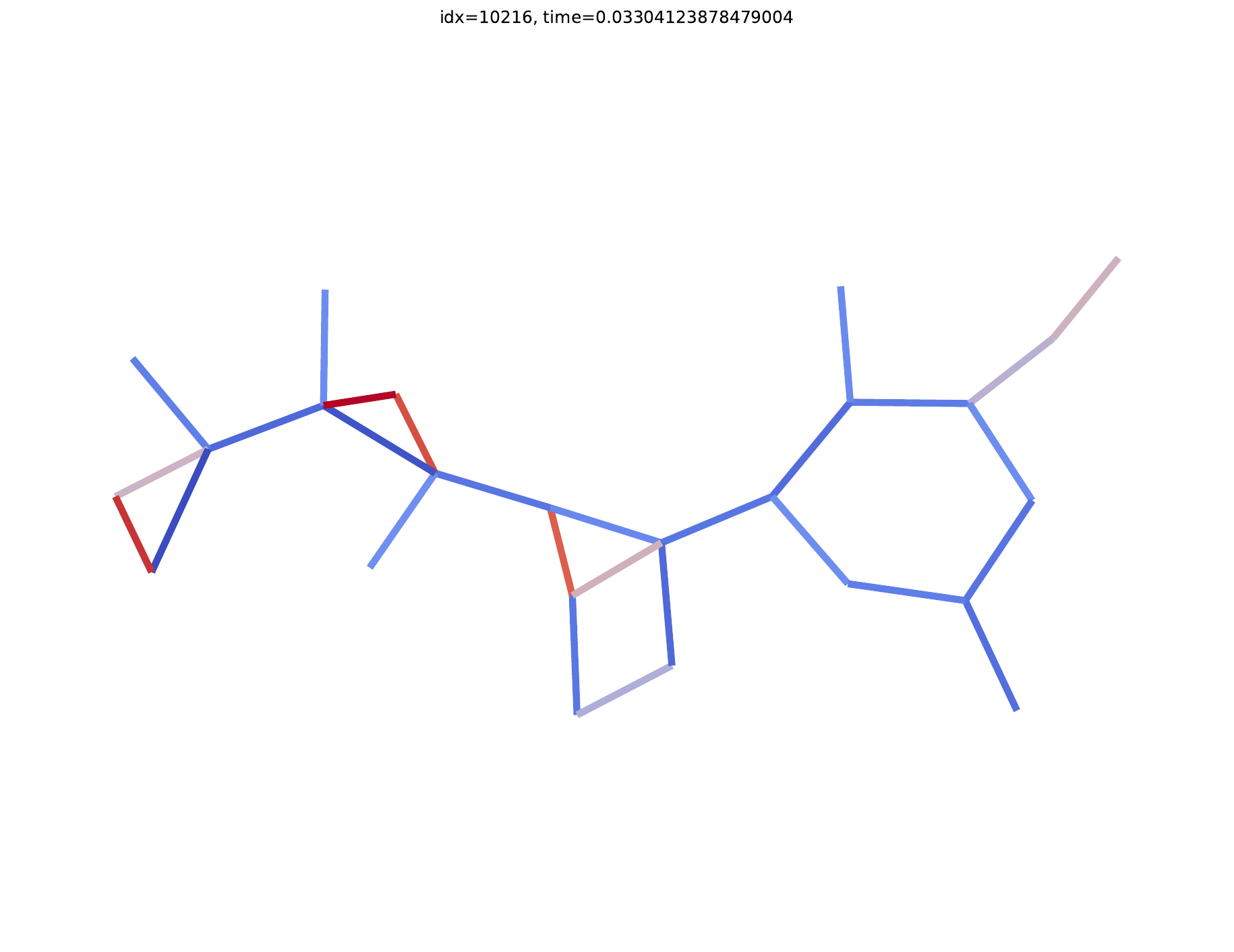} &
\imgcell{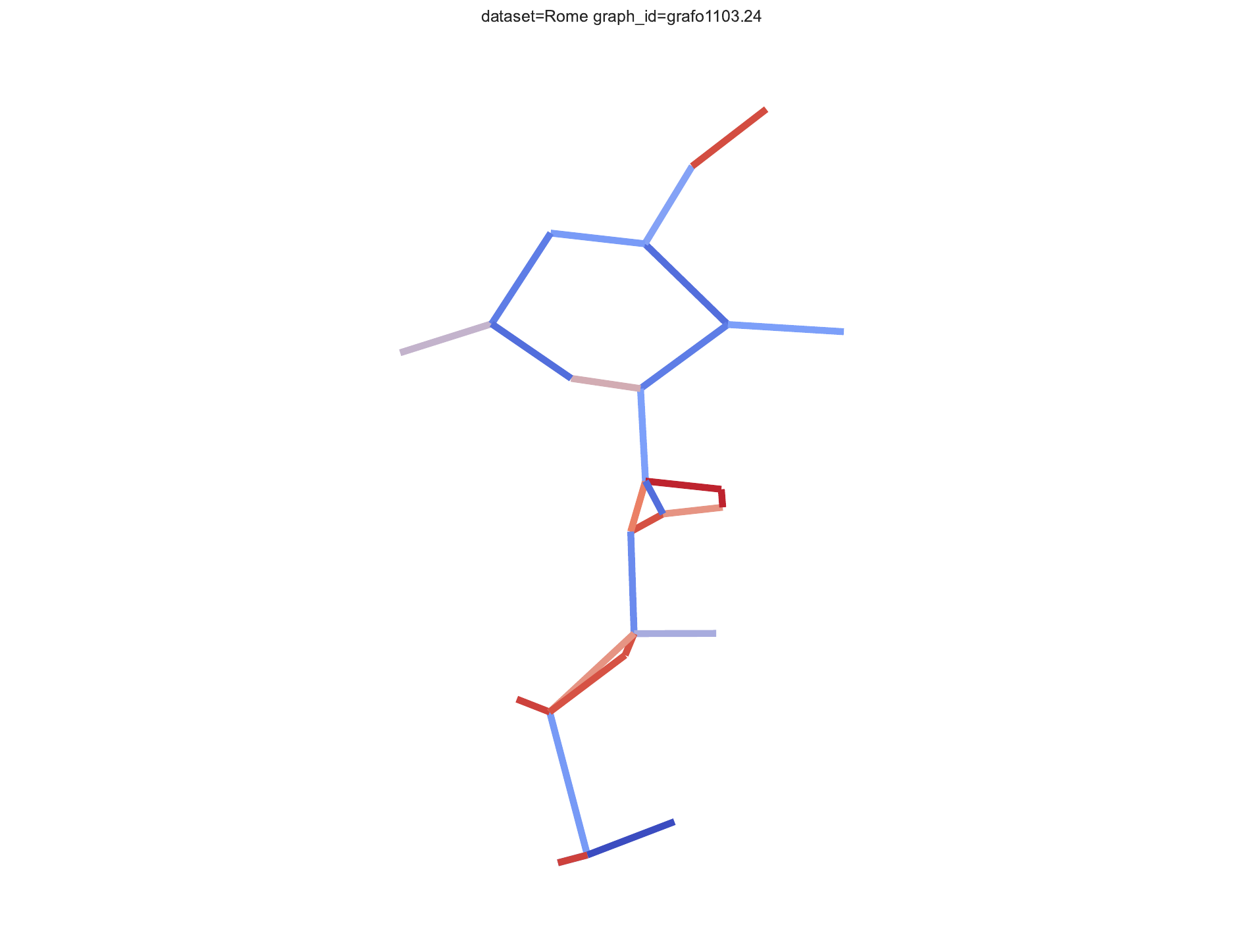} &
\imgcell{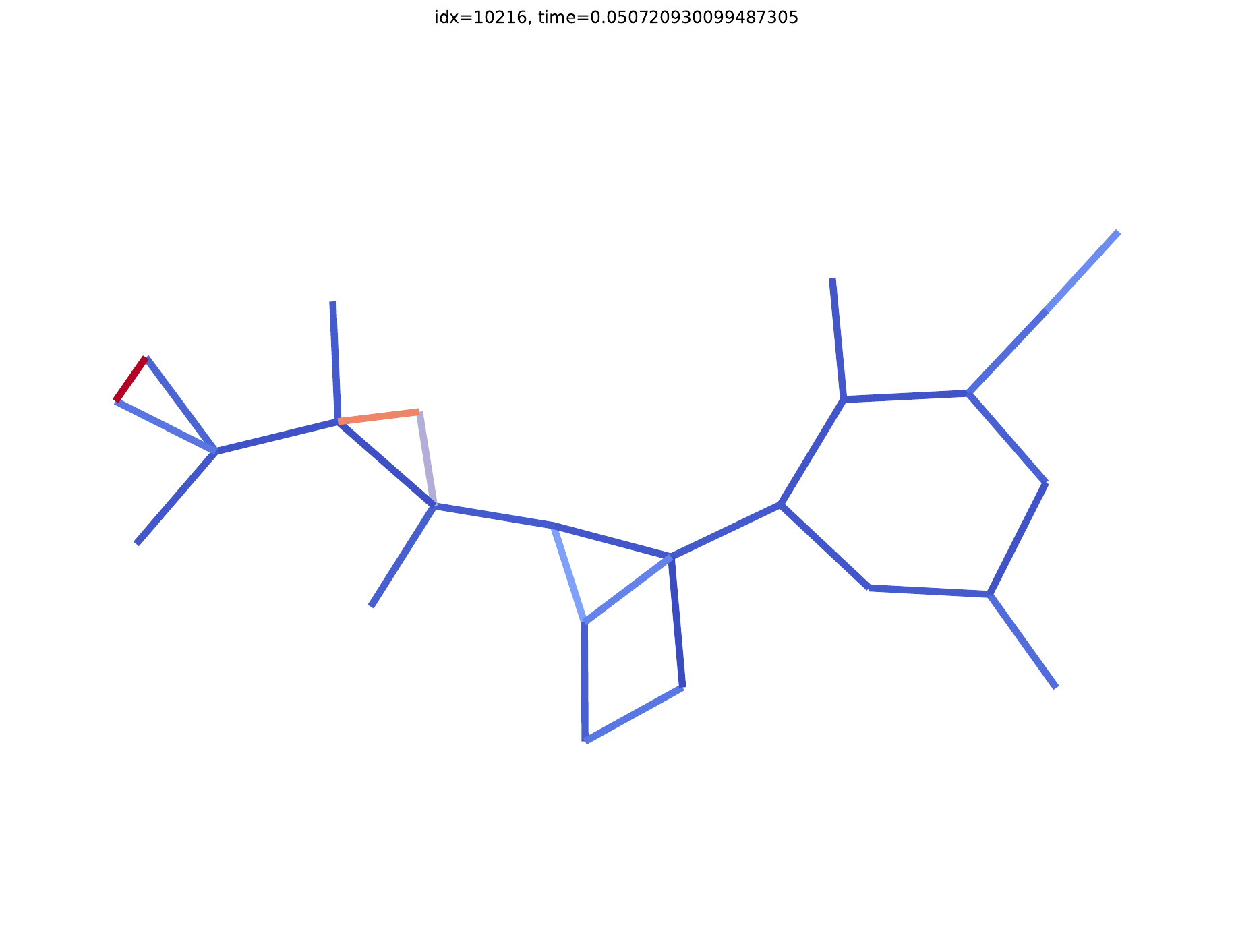} &
\imgcell{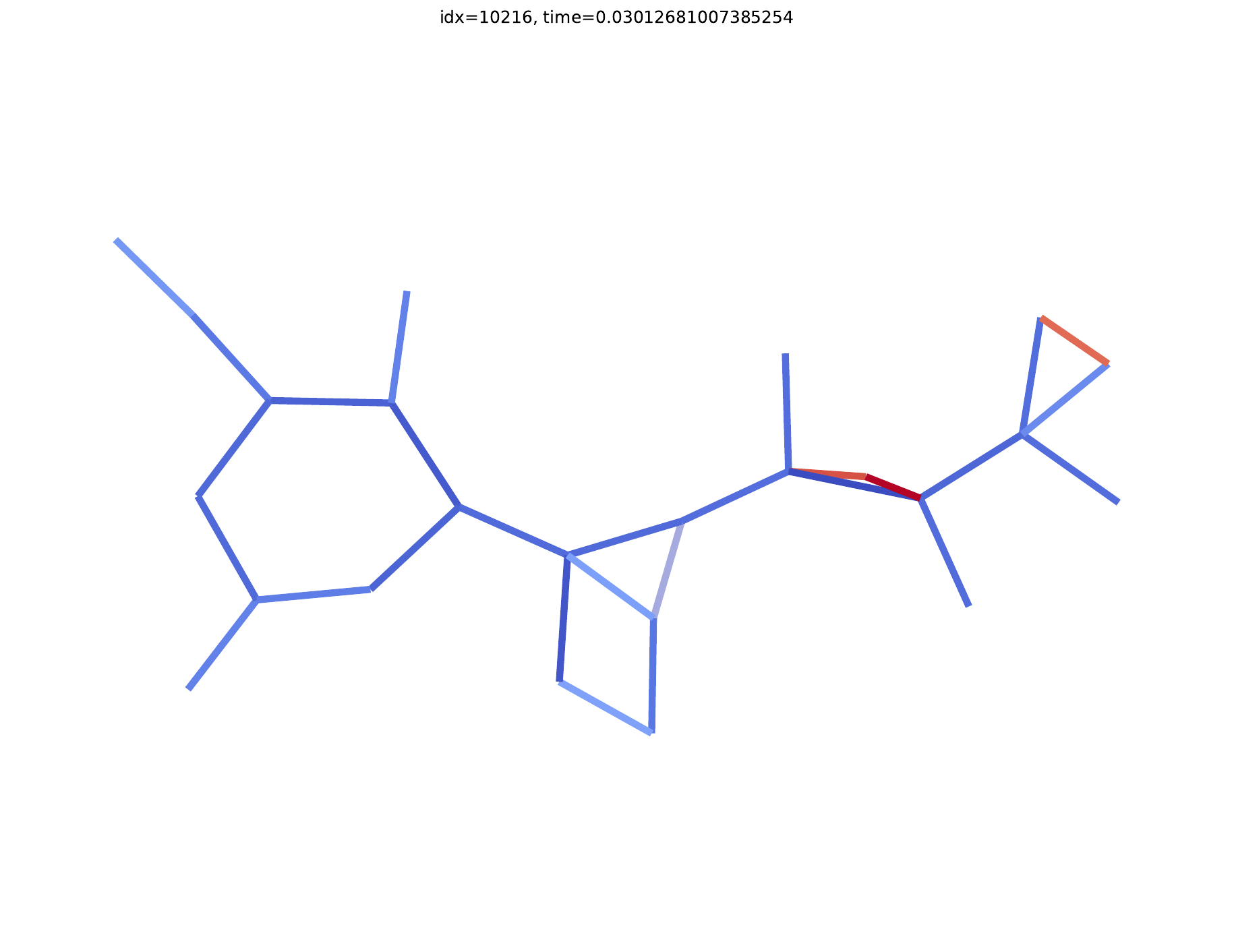} &
\imgcell{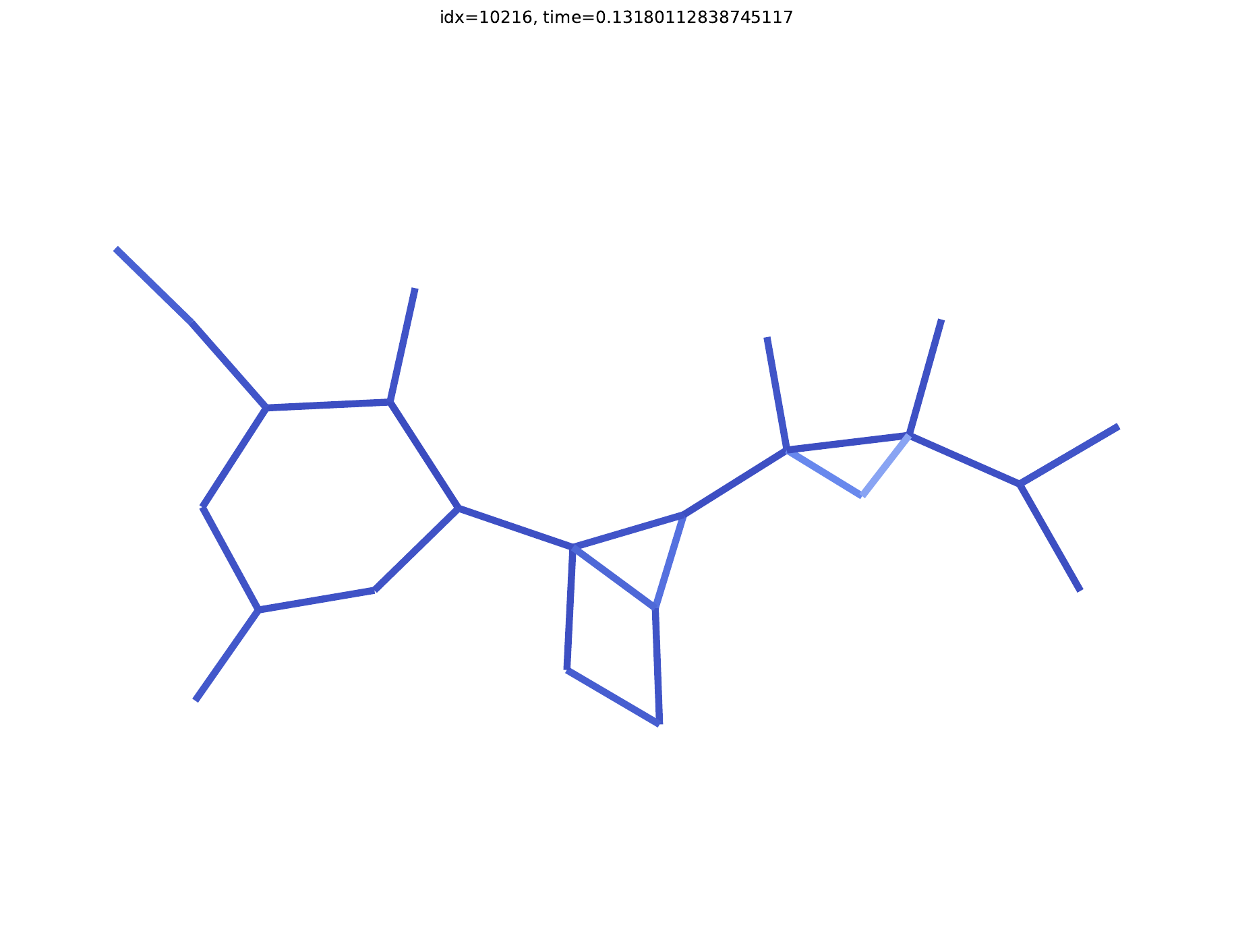} &
\imgcell{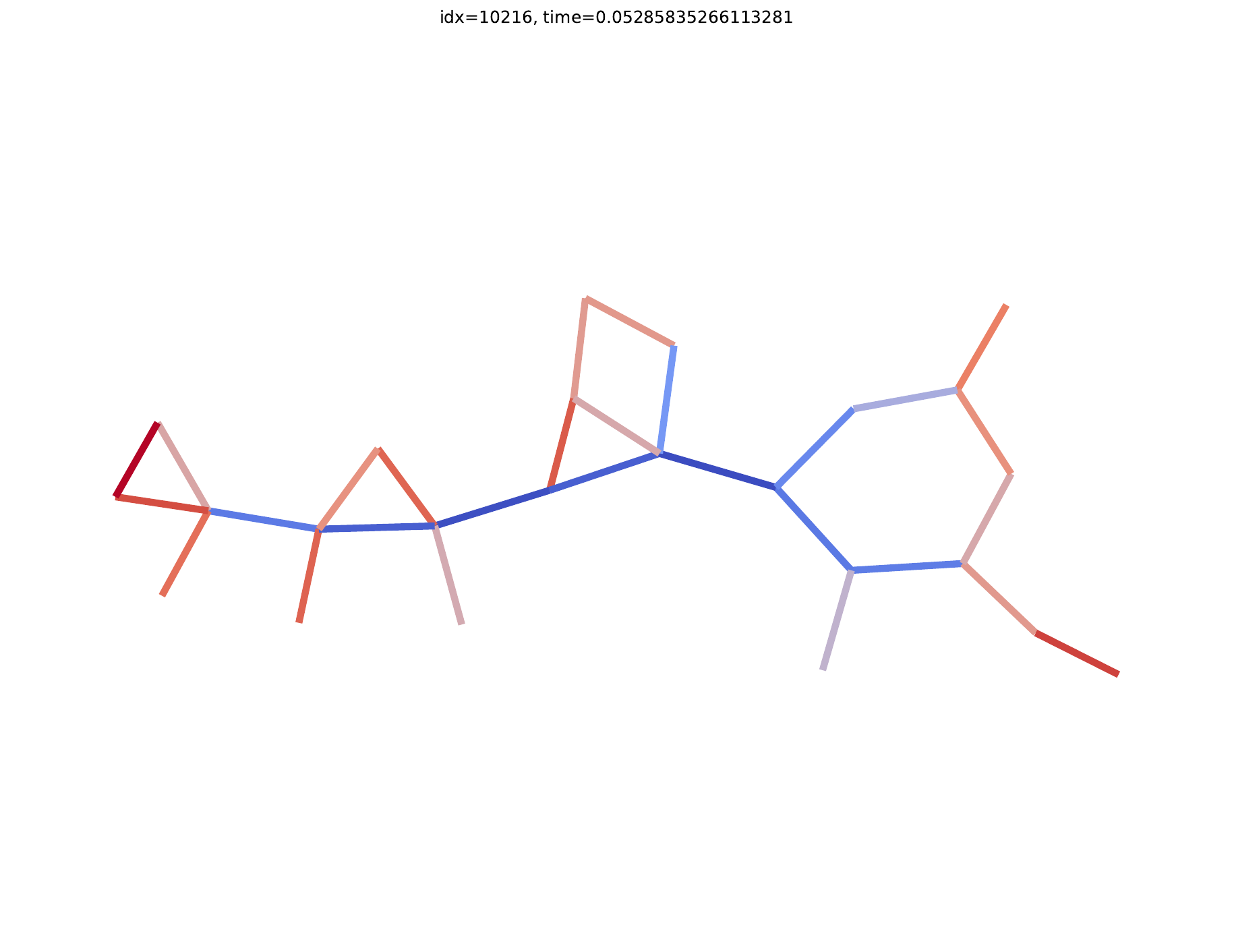} \\

&
t = 0.00s &
t = 0.28s &
t = 0.05s &
t = 0.05s &
t = 104.72s &
t = 0.04s &
t = 0.03s &
t = 0.04s &
t = 0.05s &
t = 0.03s &
t = 0.03s &
t = 0.05s \\

\end{tabular}
\captionof{figure}[]{The qualitative evaluation of 7 \modelName\ models by comparing with 5 competitive and representative benchmarks. All the graphs presented above are unseen during the training phase of \modelName. The name of the graphs with the number of nodes $N$ and the number of edges $M$ is presented in the row header. For each layout, the computation time $t$ (without including the pre-processing time) on the CPU is computed and reported in seconds. }
\label{fig:more-vis-result3}
\end{table*}

\begin{table*}[ht!]
\setlength{\tabcolsep}{0pt}
\renewcommand{\arraystretch}{0}
\fontsize{6}{6}\selectfont
\centering
\begin{tabular}{ c|ccccc|ccccccc }
    \bfseries{\thead{Graph}} & \multicolumn{5}{c|}{\thead{Benchmark Methods}} & \multicolumn{6}{c}{\thead{SmartGD}}\\
    & \bfseries{SGD2}
    & \bfseries{PMDS}
    & \bfseries{FA2}
    & \bfseries{DeepGD}
    & \makecell{\bfseries GD2\\\relax[Stress+Xing]}
    & \makecell{\bfseries SmartGD\\\relax[Stress]}
    & \makecell{\bfseries SmartGD\\\relax[Xing]}
    & \makecell{\bfseries SmartGD\\\relax[Shape]}
    & \makecell{\bfseries SmartGD\\\relax[XAngle]}
    & \makecell{\bfseries SmartGD\\\relax[Stress+Xing]}
    & \makecell{\bfseries SmartGD\\\relax[Stress+XAngle]}
    & \makecell{\bfseries SmartGD\\\relax[7-Aesthetics]}
    \rule[-1ex]{0pt}{0ex} \\ \hline

\makecell{\bfseries grafo428.19\\N = 12\\M = 12} &
\imgcell{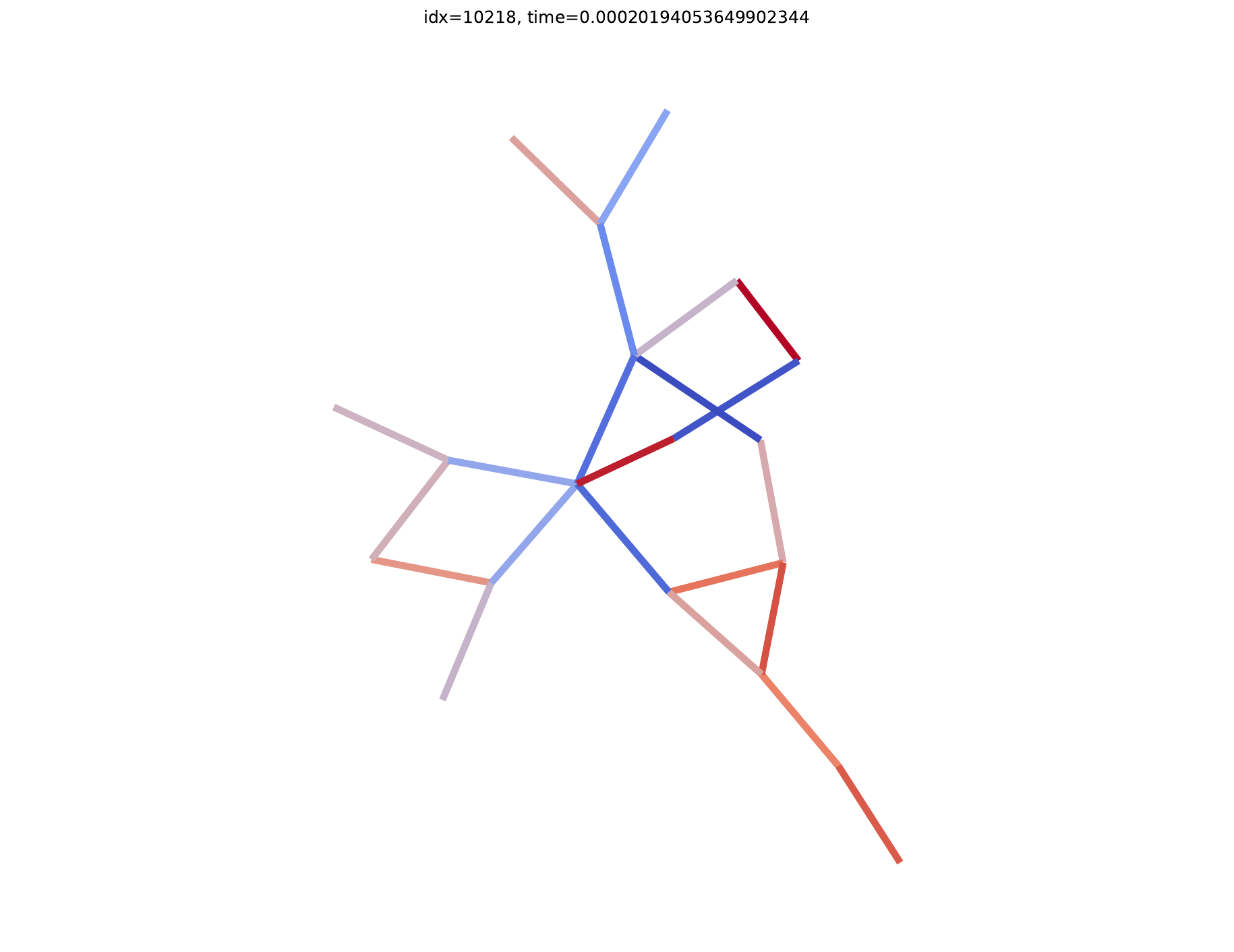} &
\imgcell{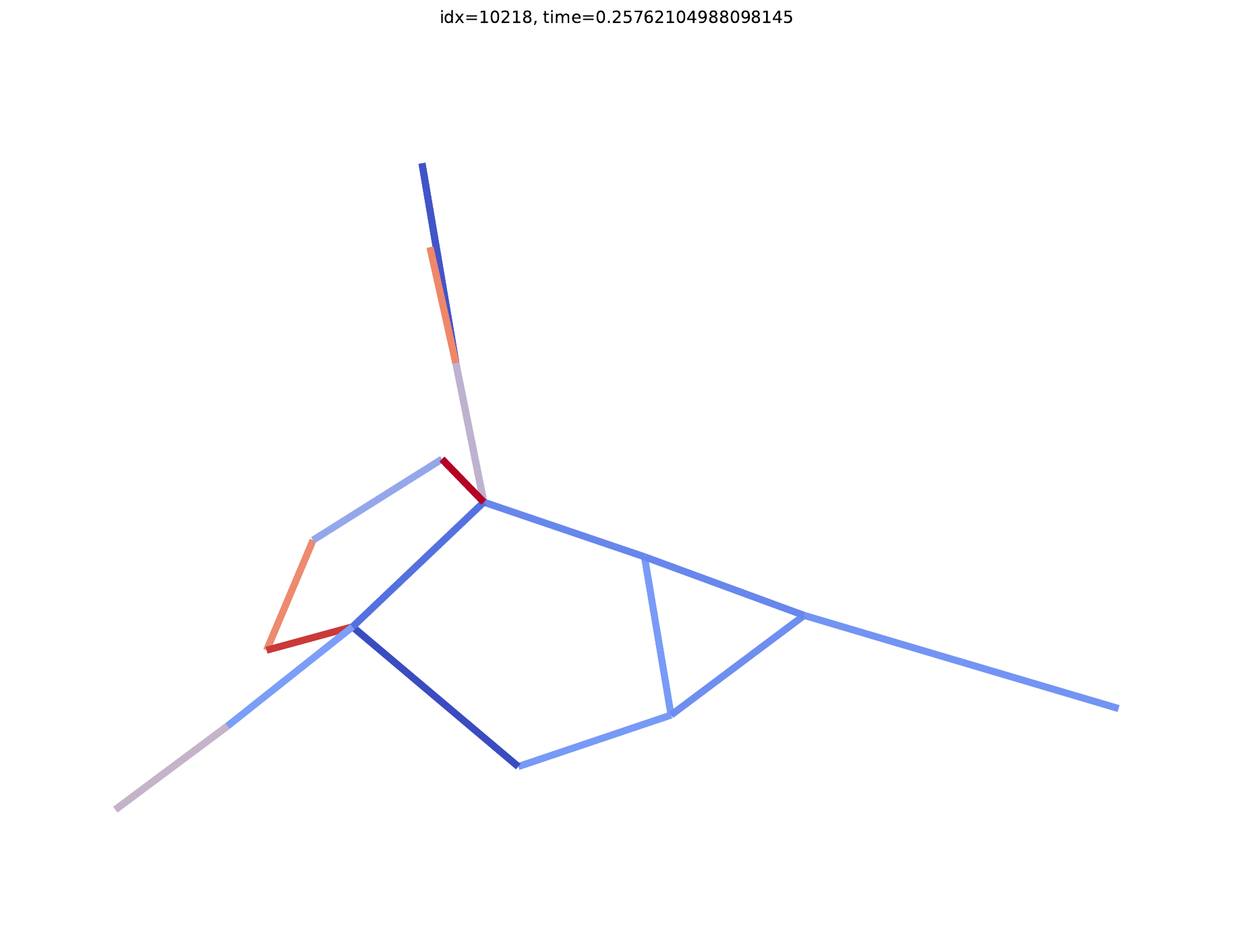} &
\imgcell{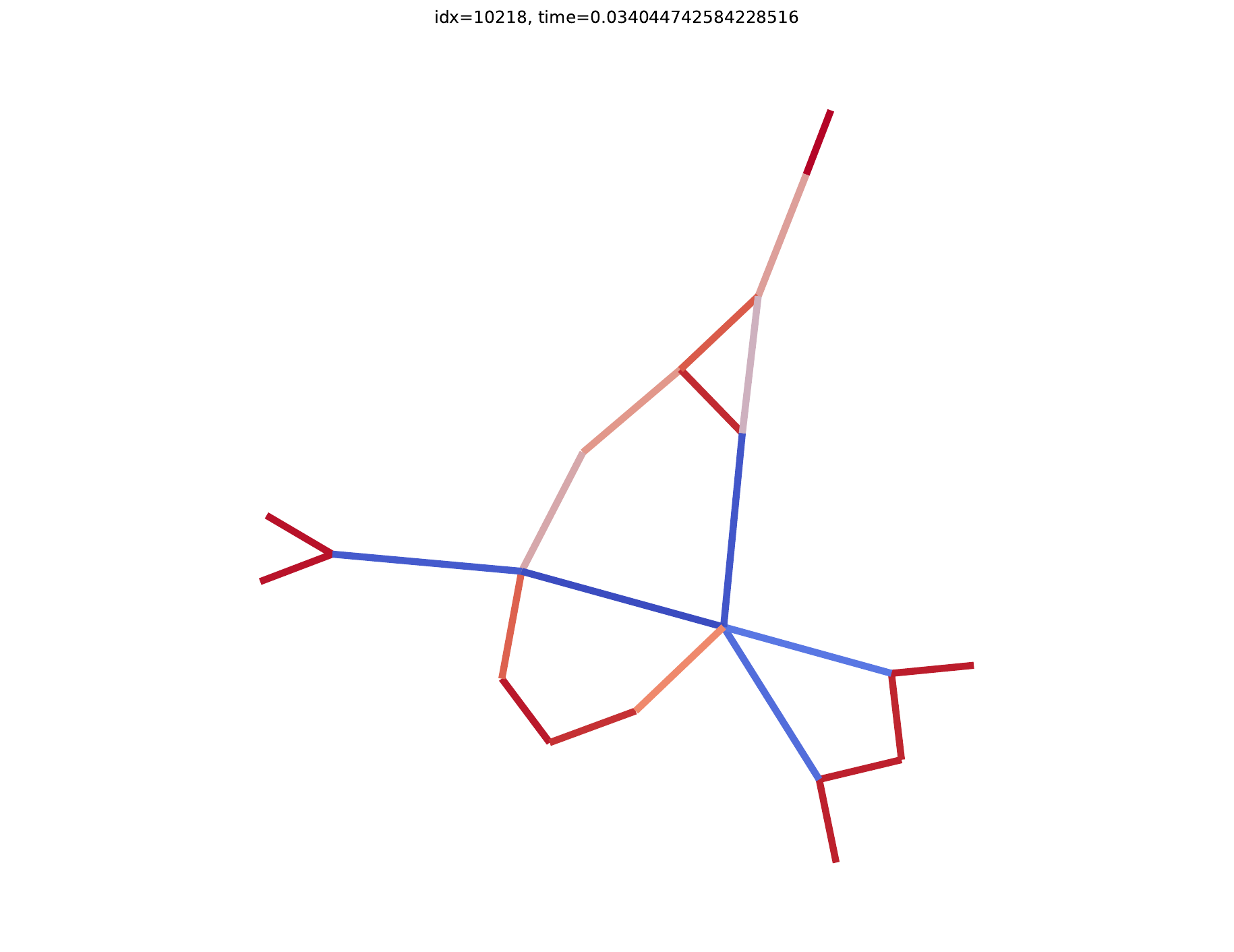} &
\imgcell{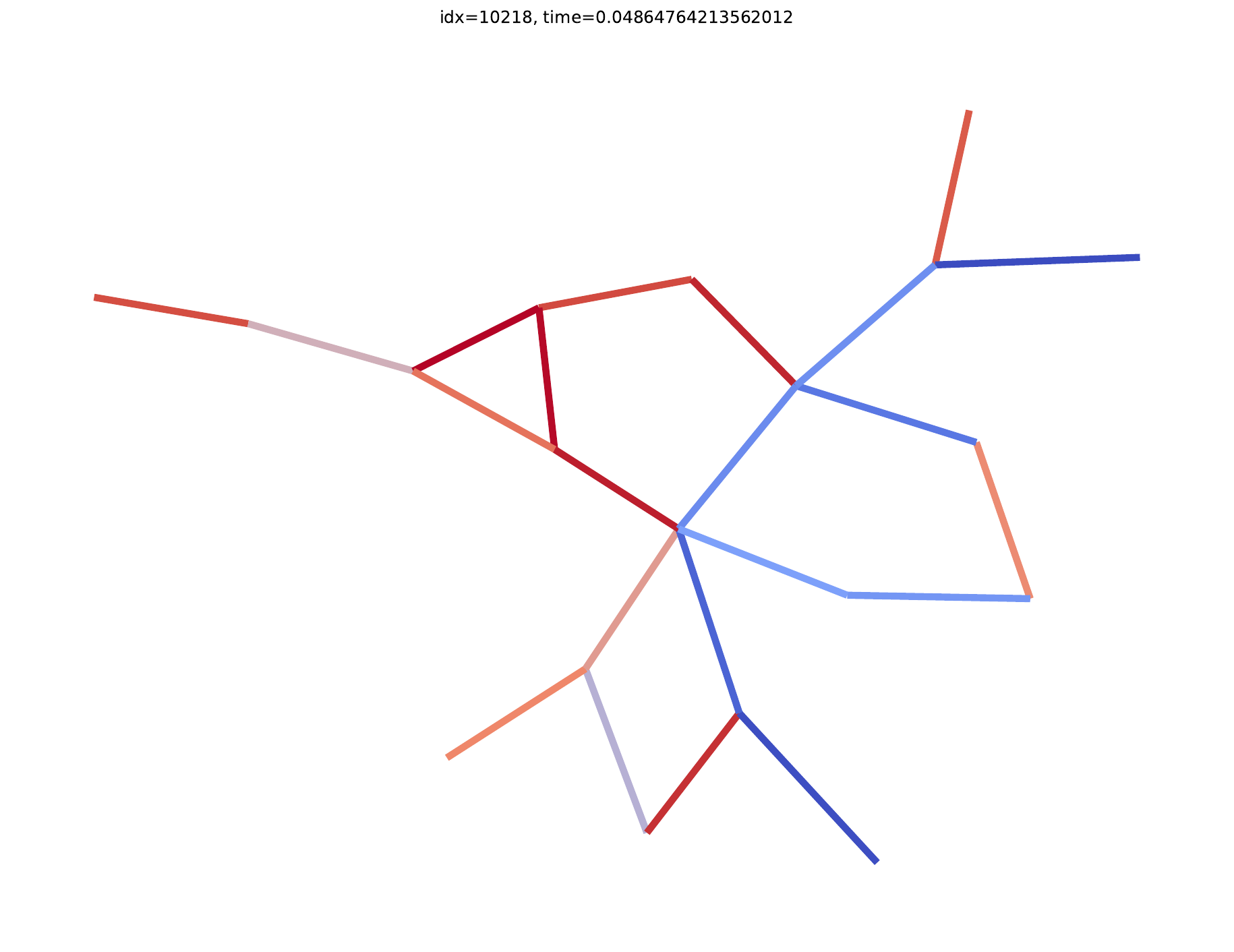} &
\imgcell{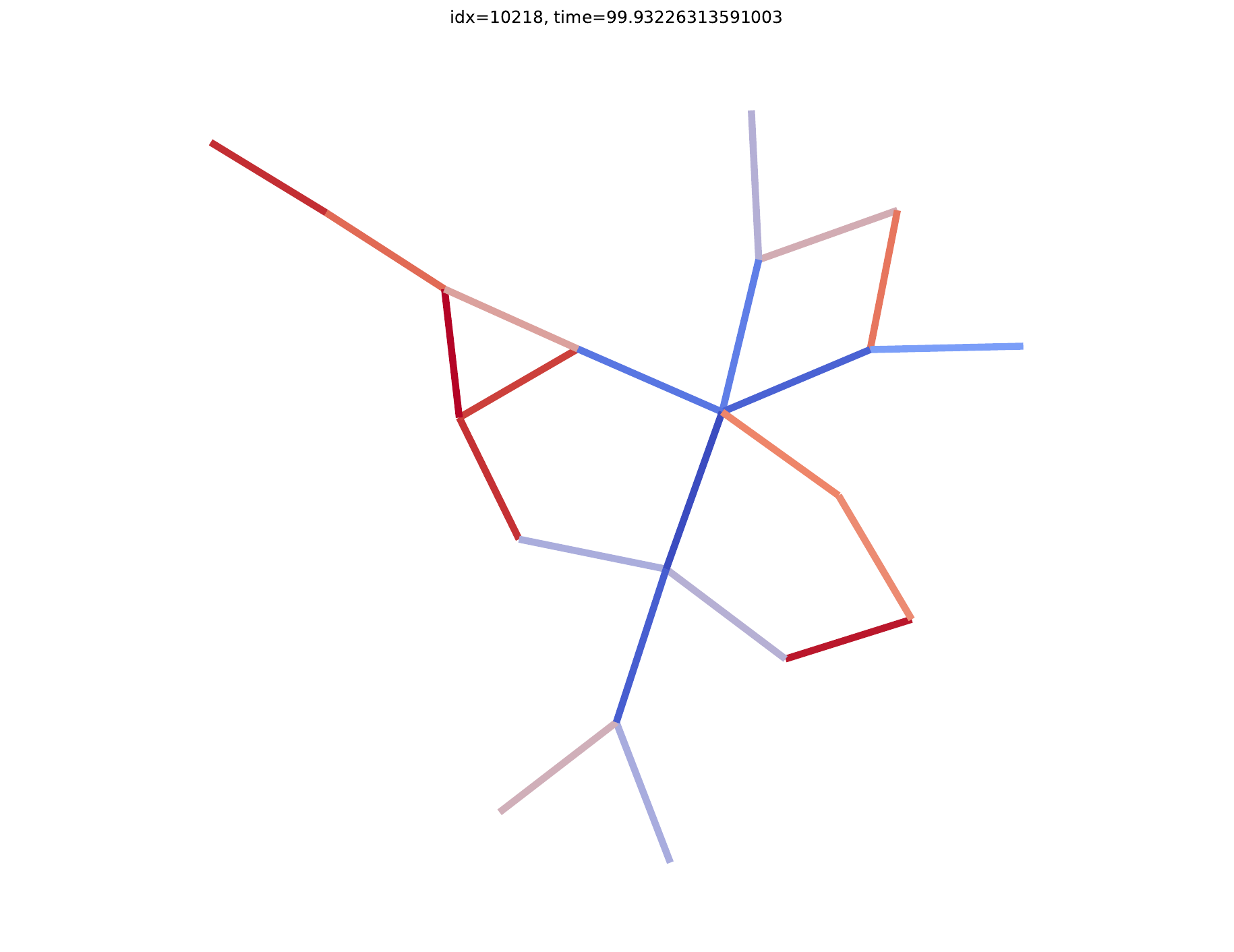} &
\imgcell{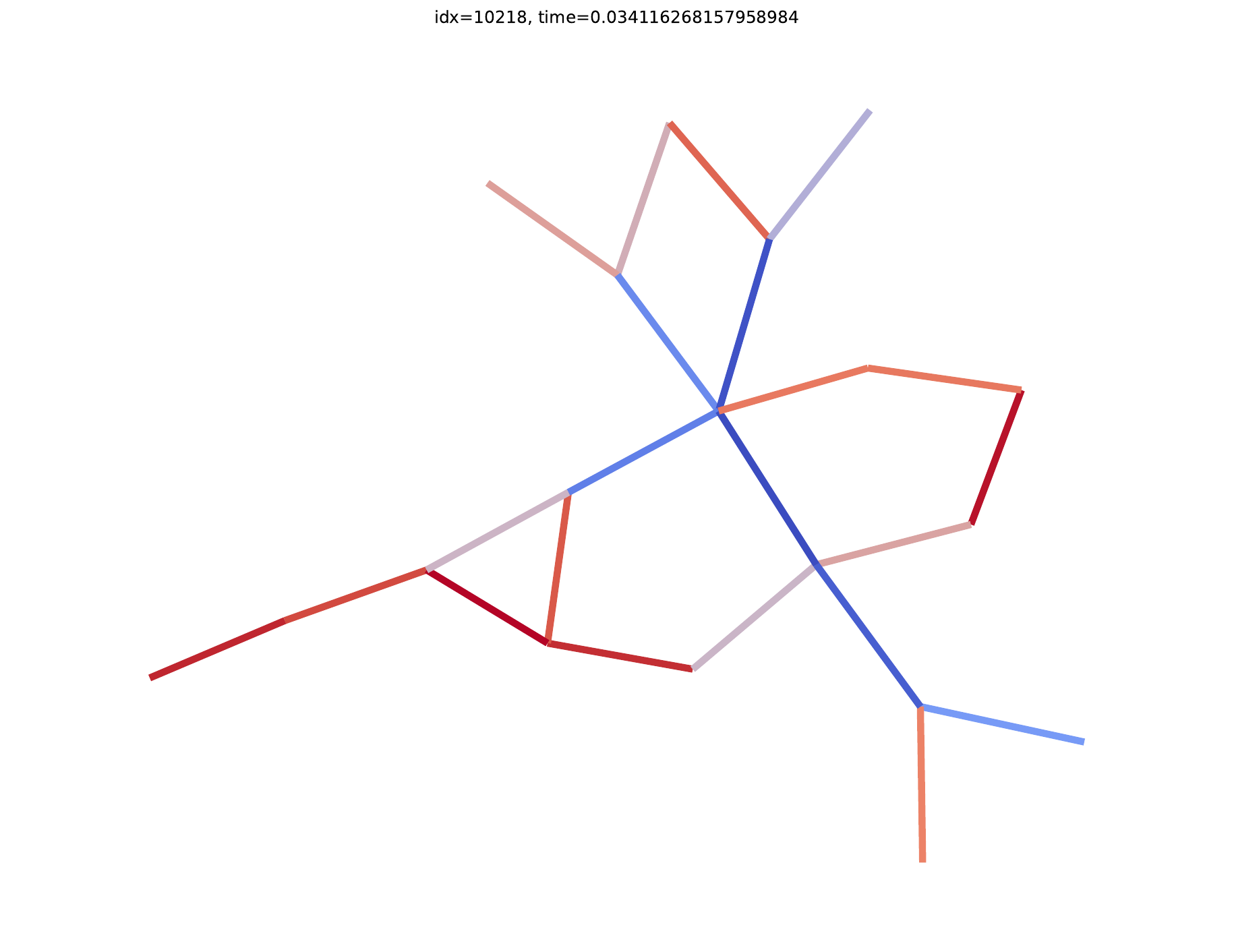} &
\imgcell{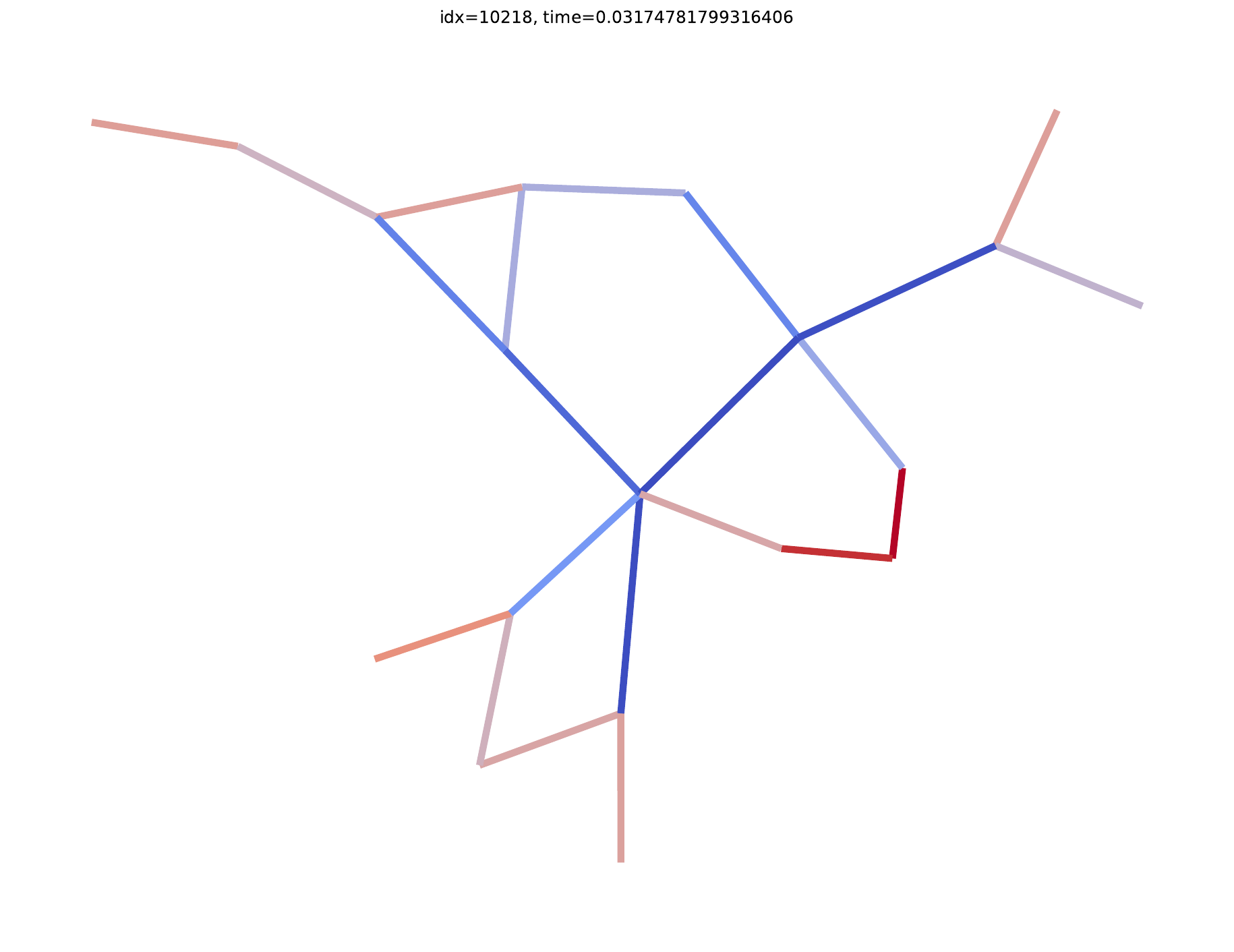} &
\imgcell{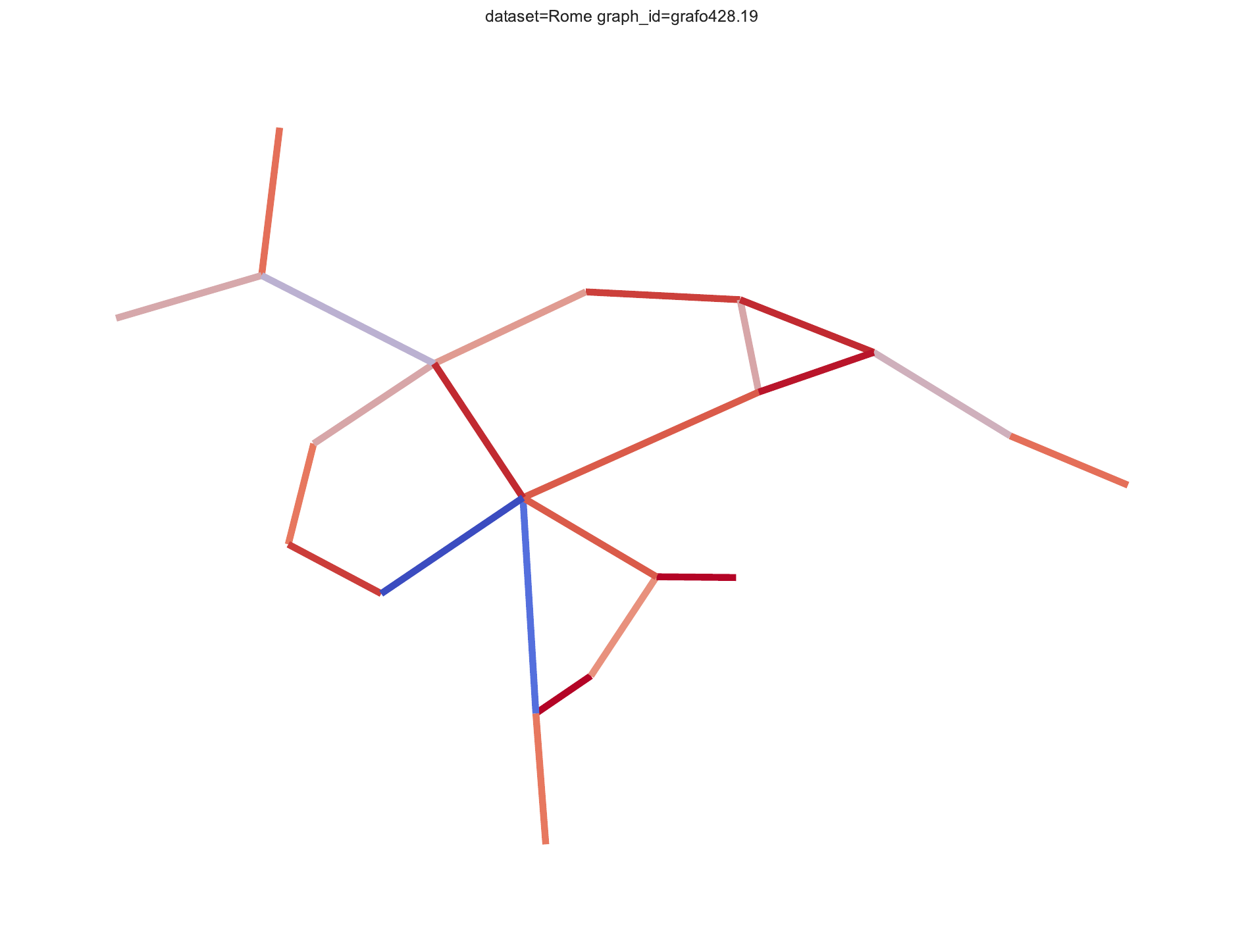} &
\imgcell{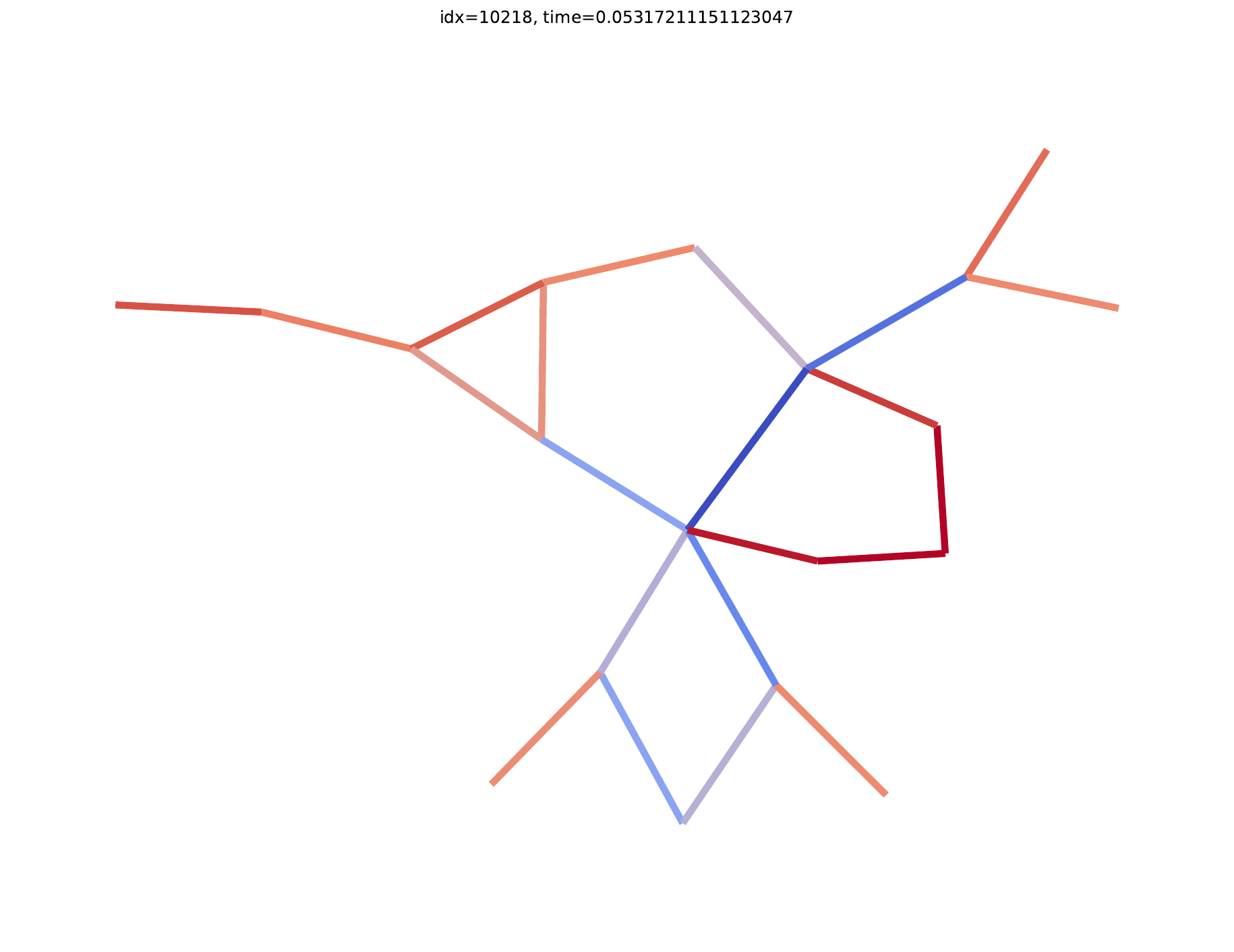} &
\imgcell{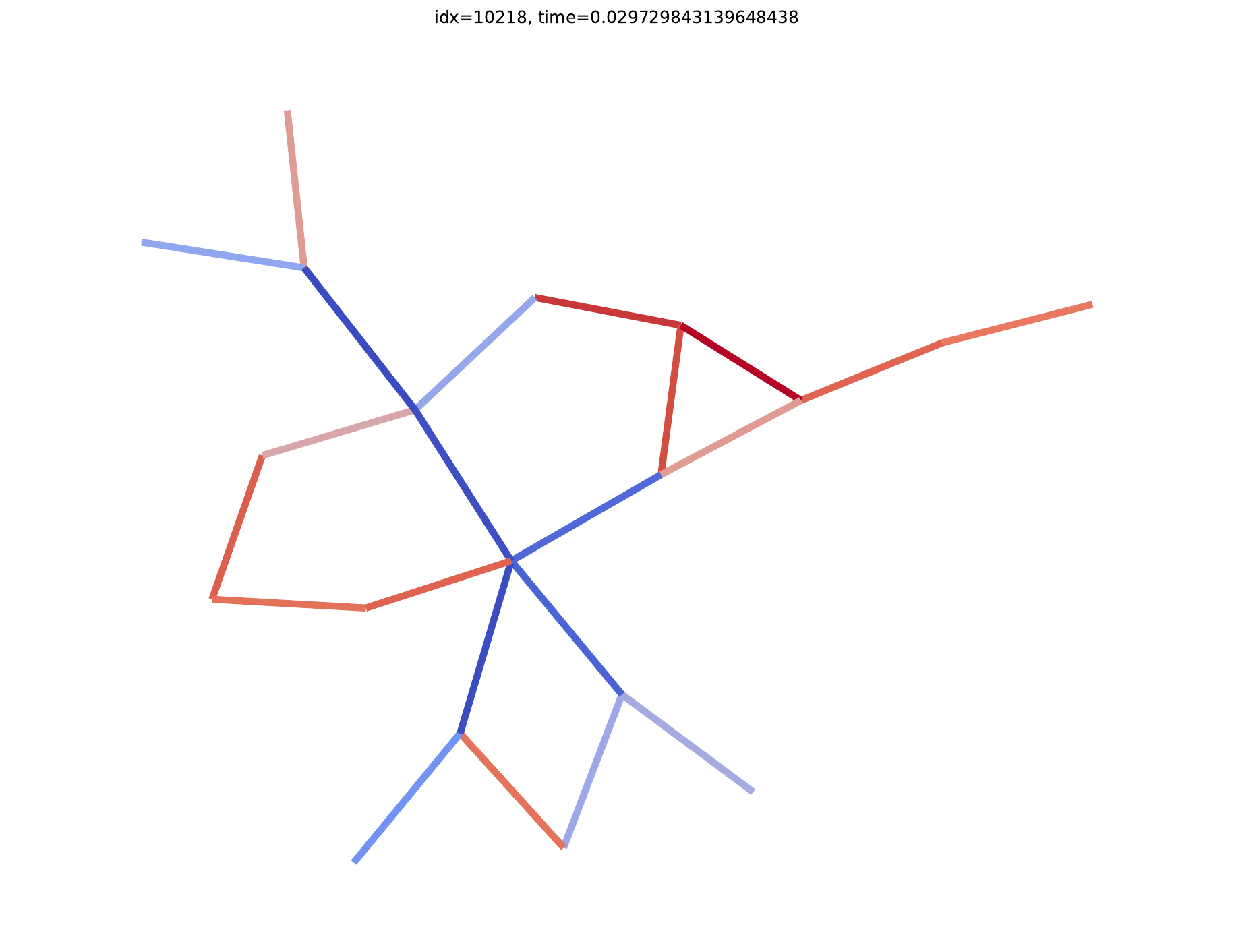} &
\imgcell{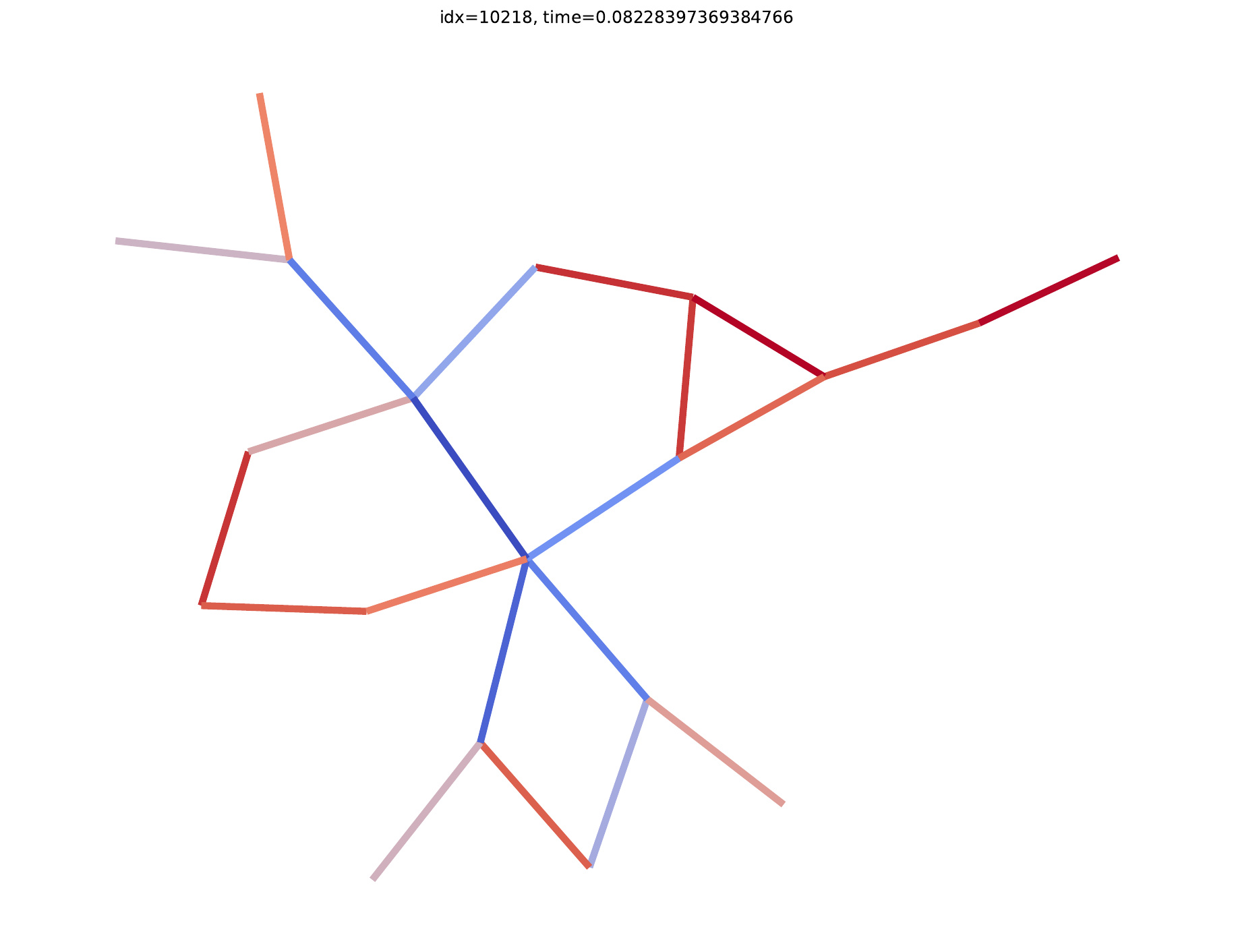} &
\imgcell{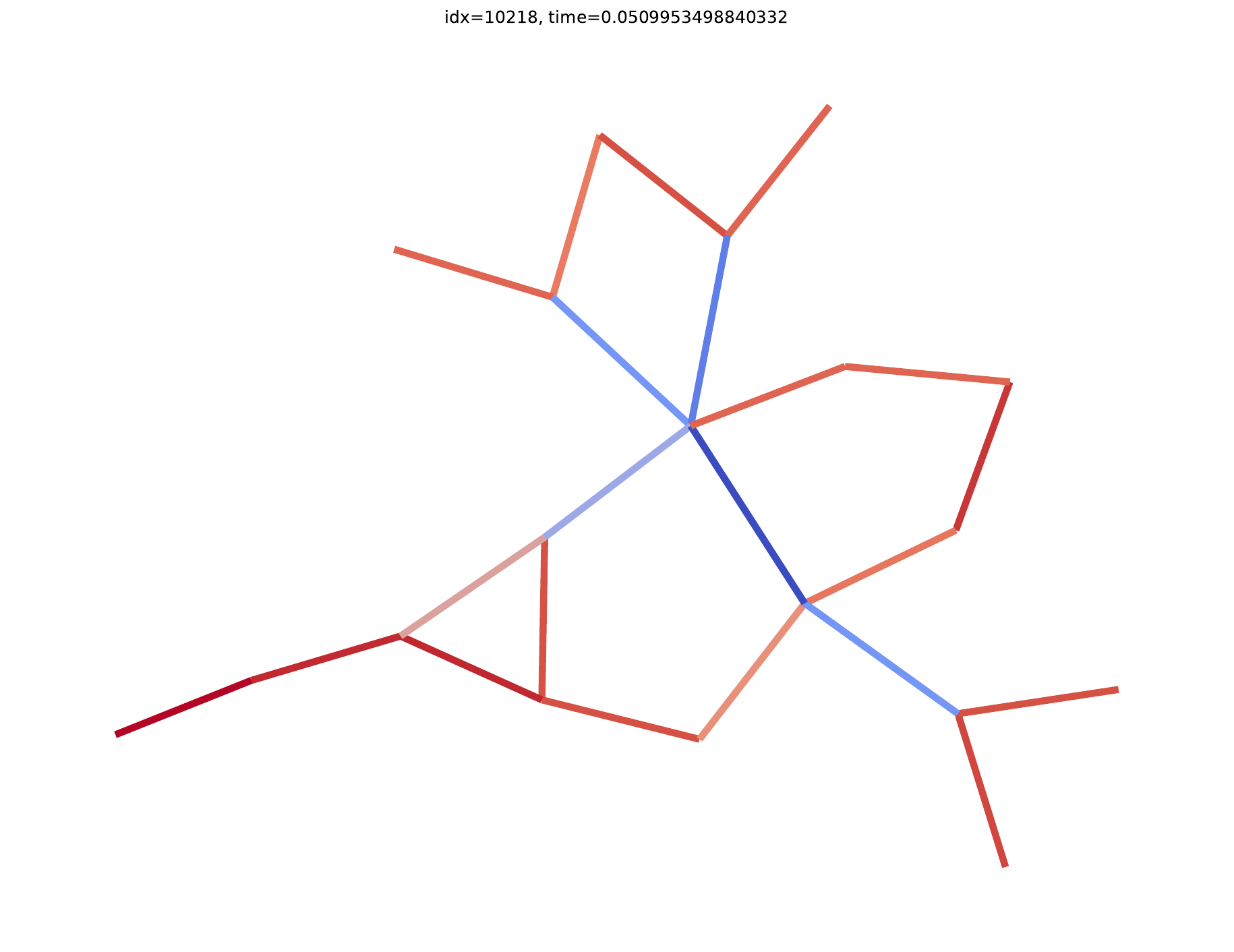} \\

&
t = 0.00s &
t = 0.26s &
t = 0.03s &
t = 0.05s &
t = 99.93s &
t = 0.03s &
t = 0.03s &
t = 0.03s &
t = 0.05s &
t = 0.03s &
t = 0.04s &
t = 0.05s \\

\makecell{\bfseries grafo2778.80\\N = 36\\M = 43} &
\imgcell{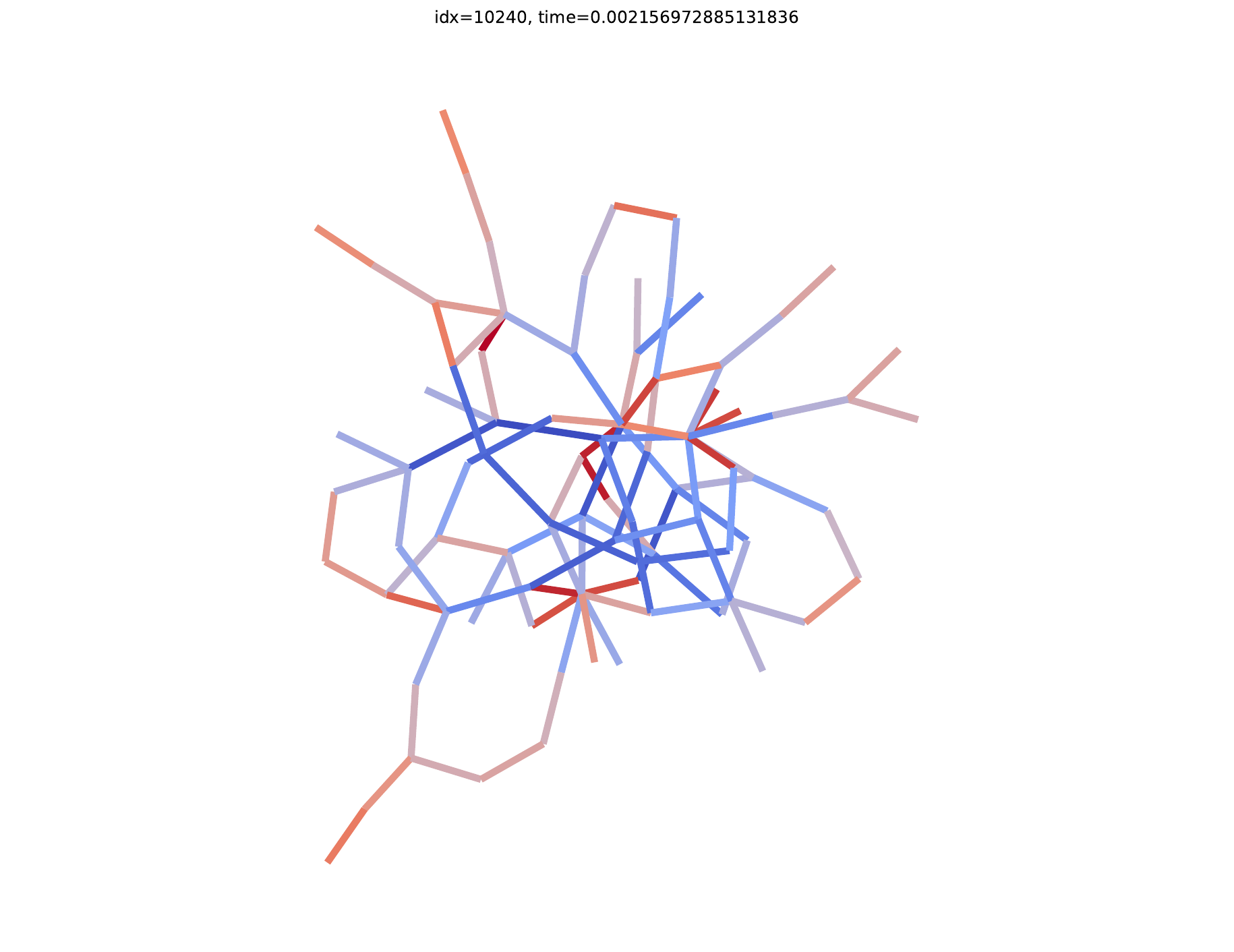} &
\imgcell{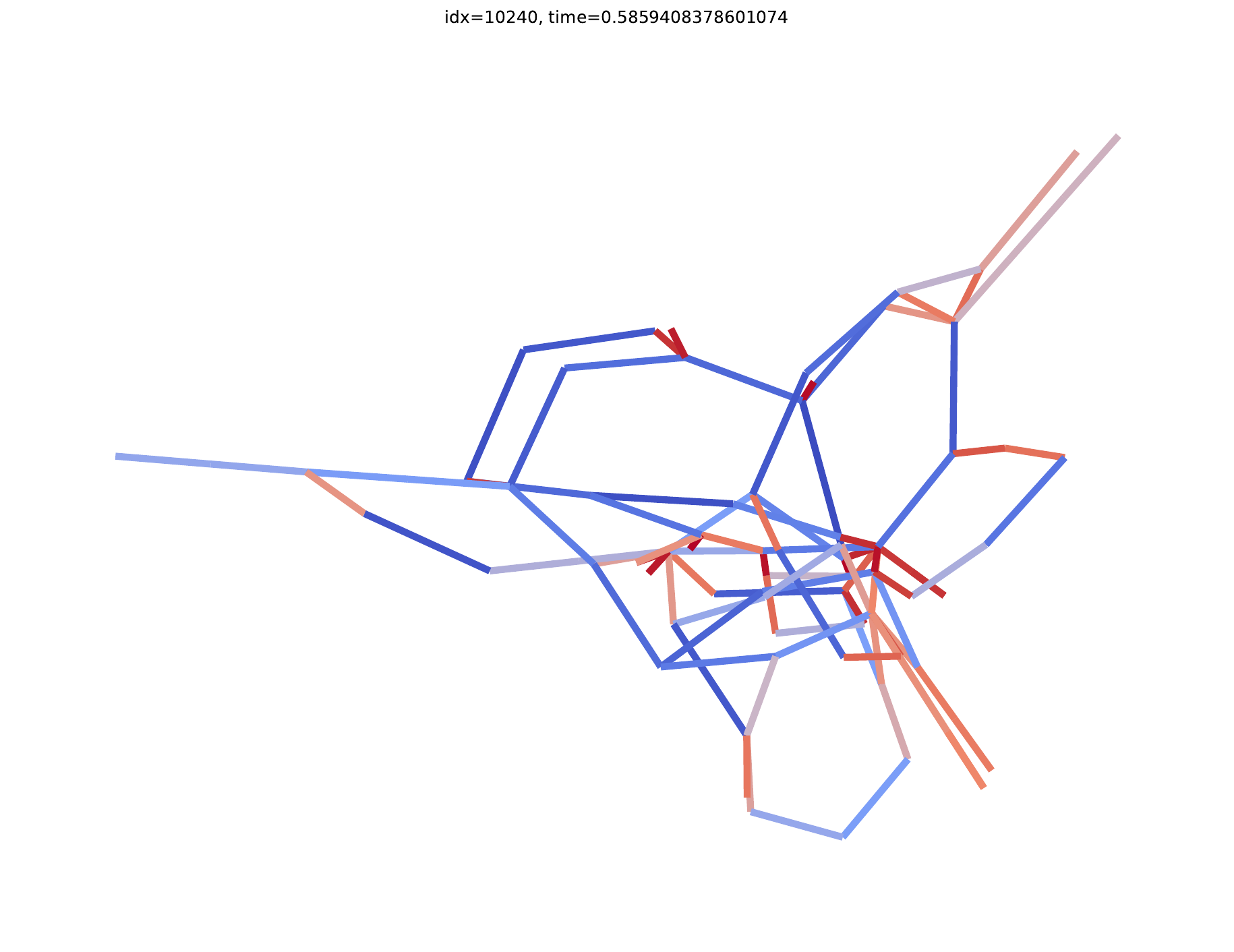} &
\imgcell{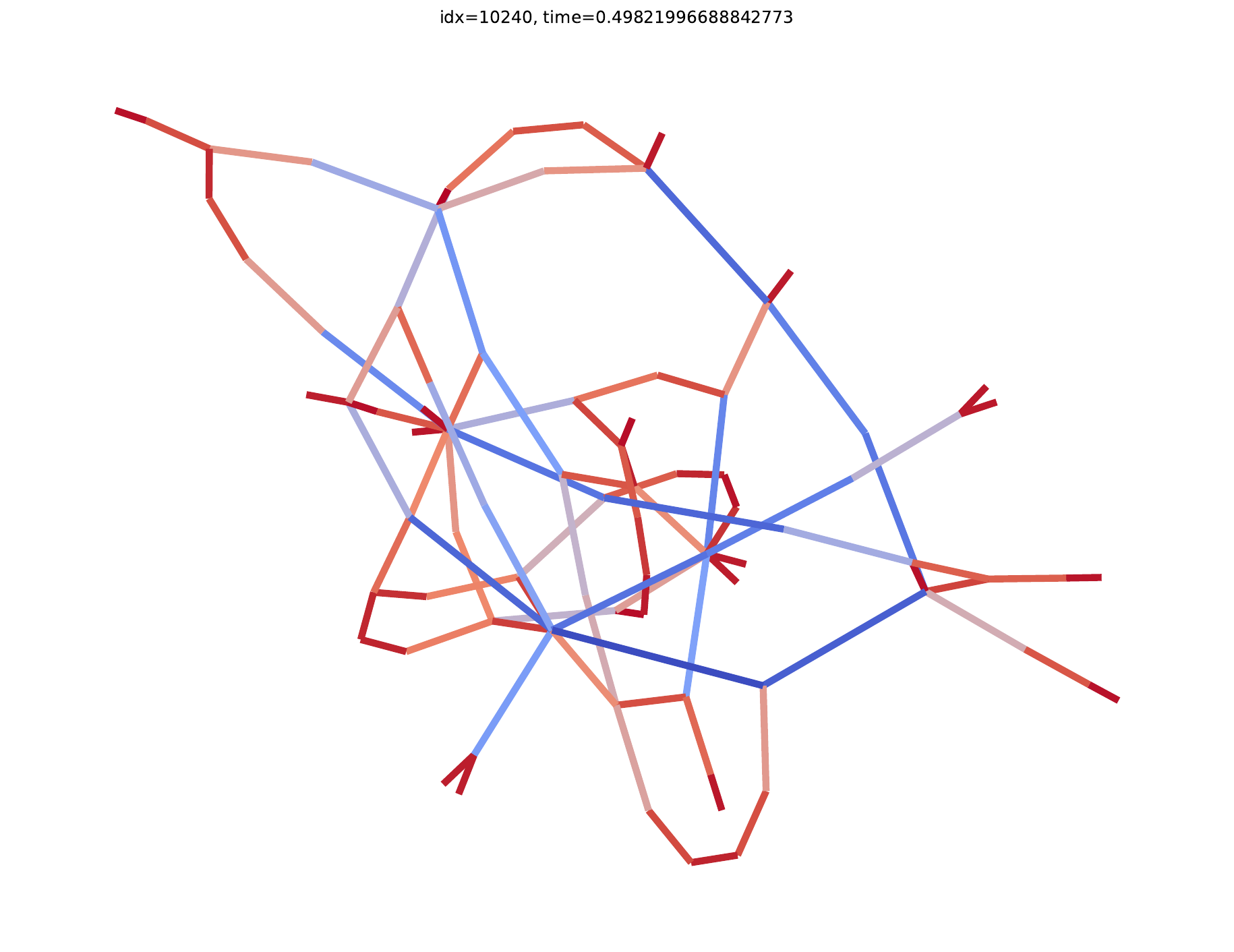} &
\imgcell{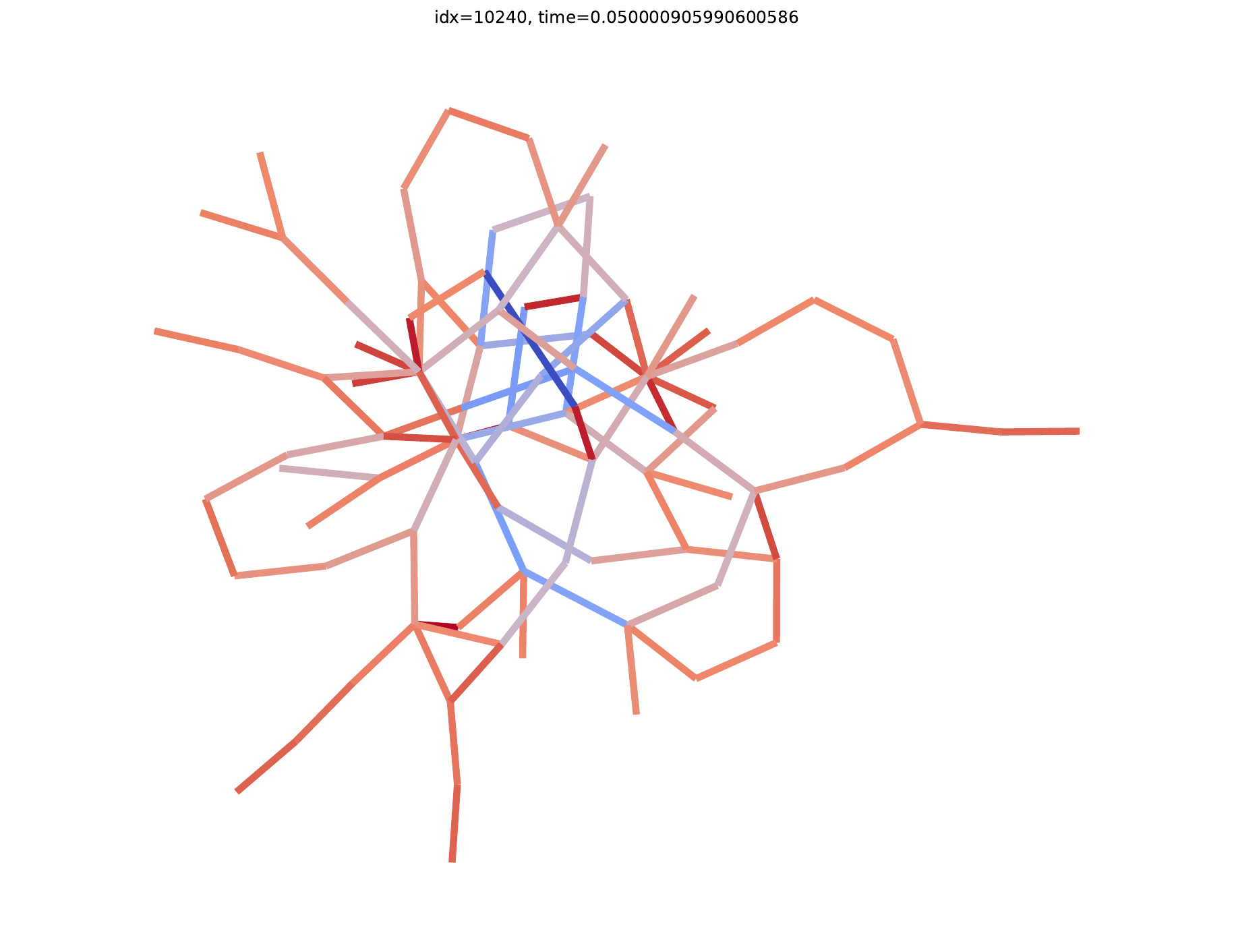} &
\imgcell{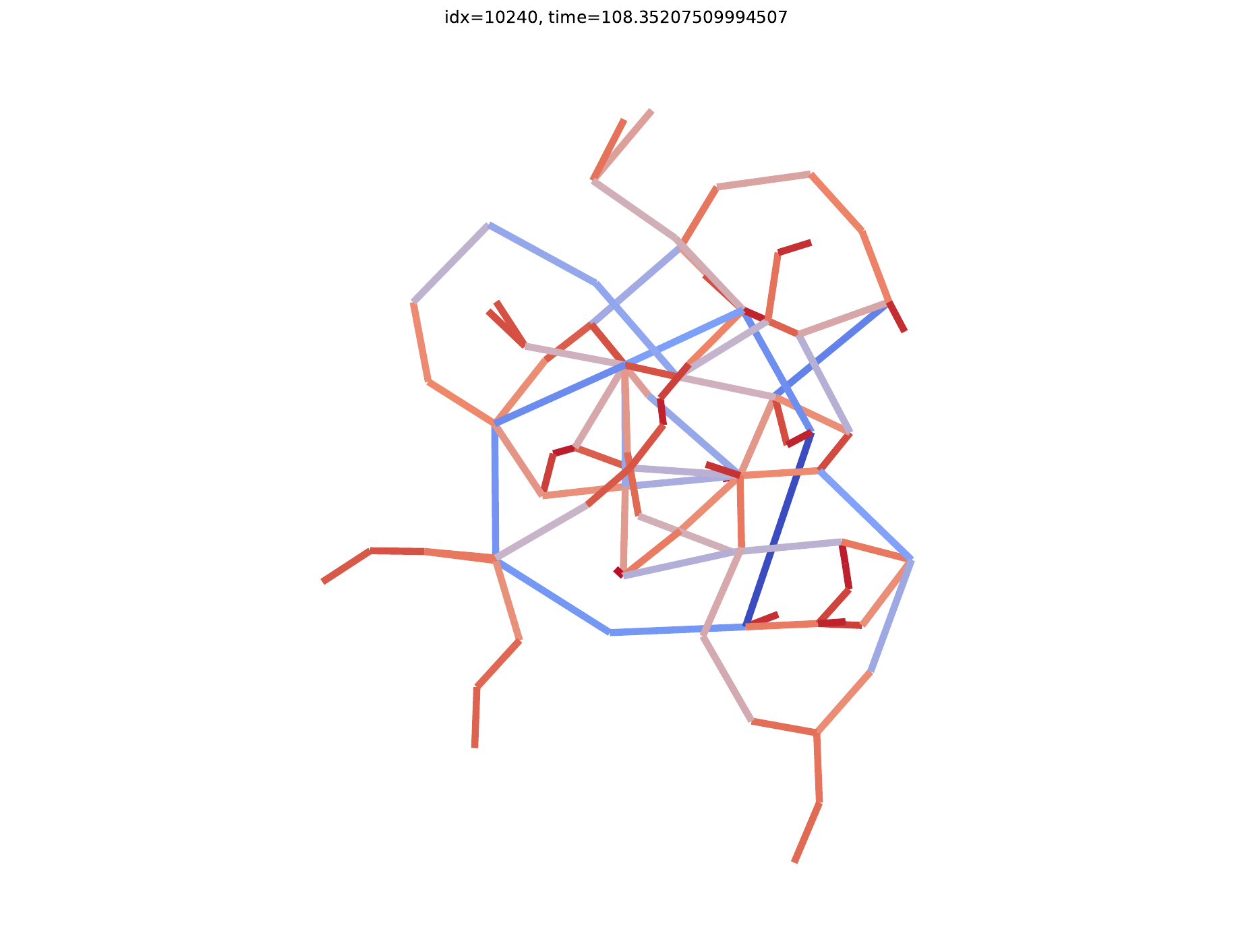} &
\imgcell{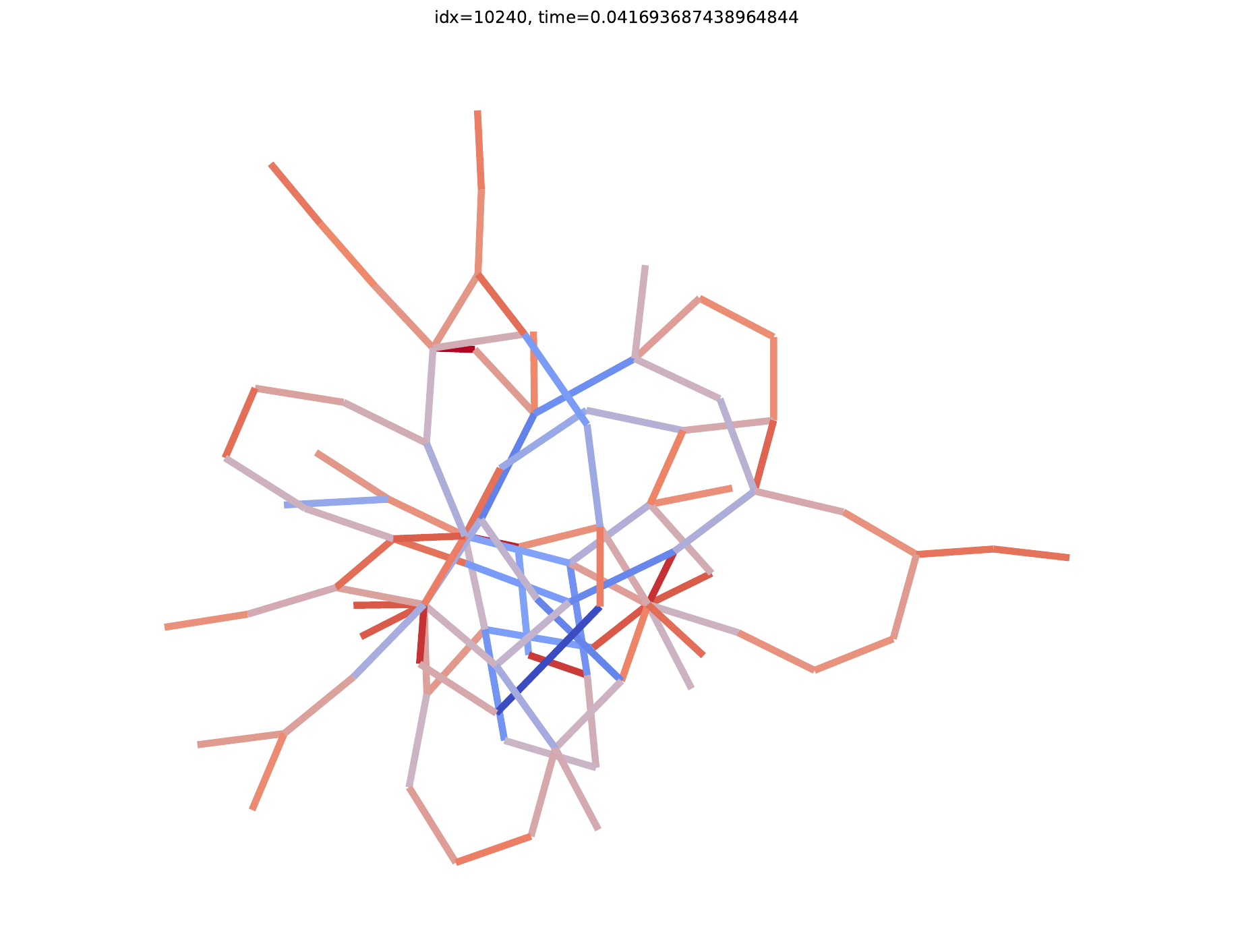} &
\imgcell{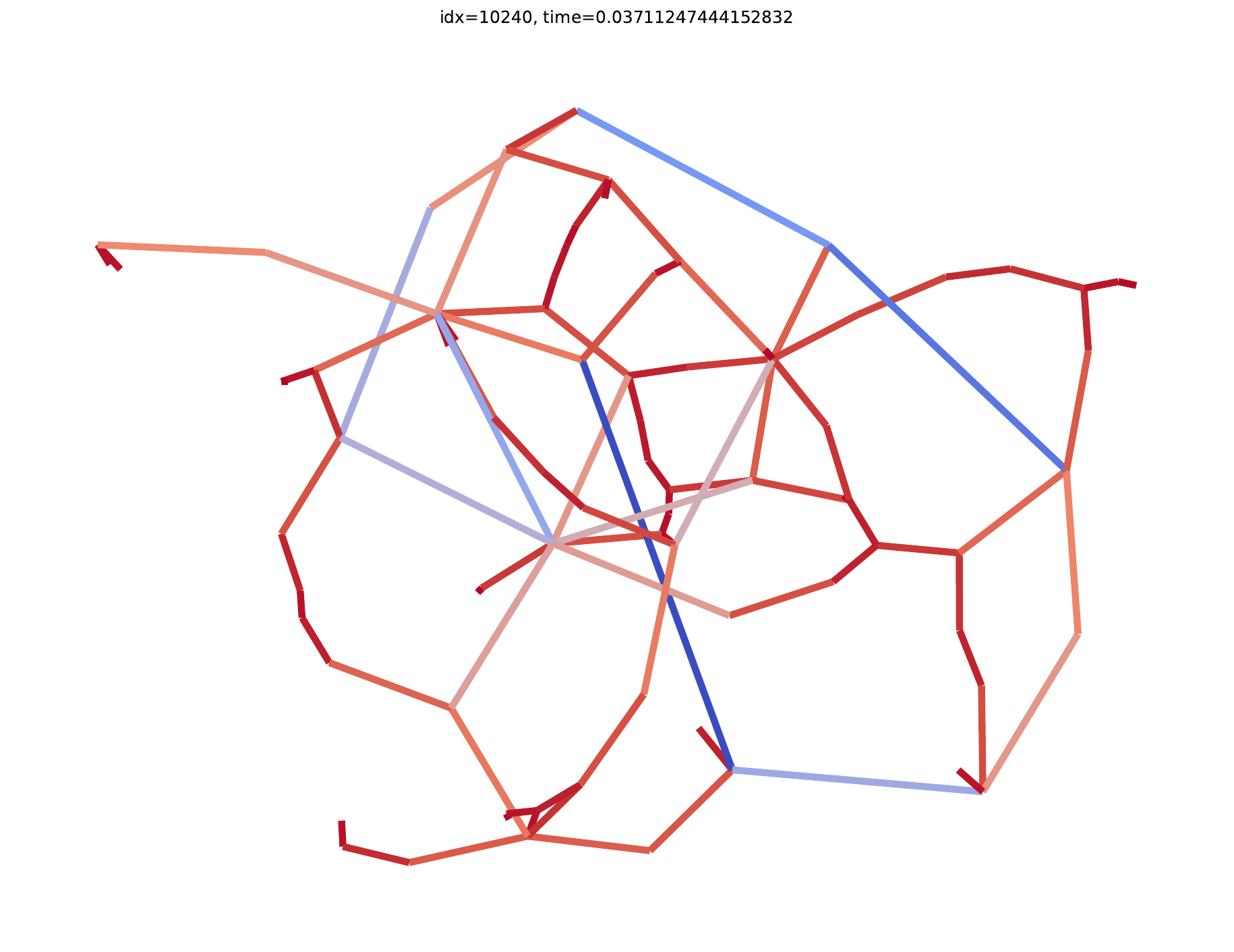} &
\imgcell{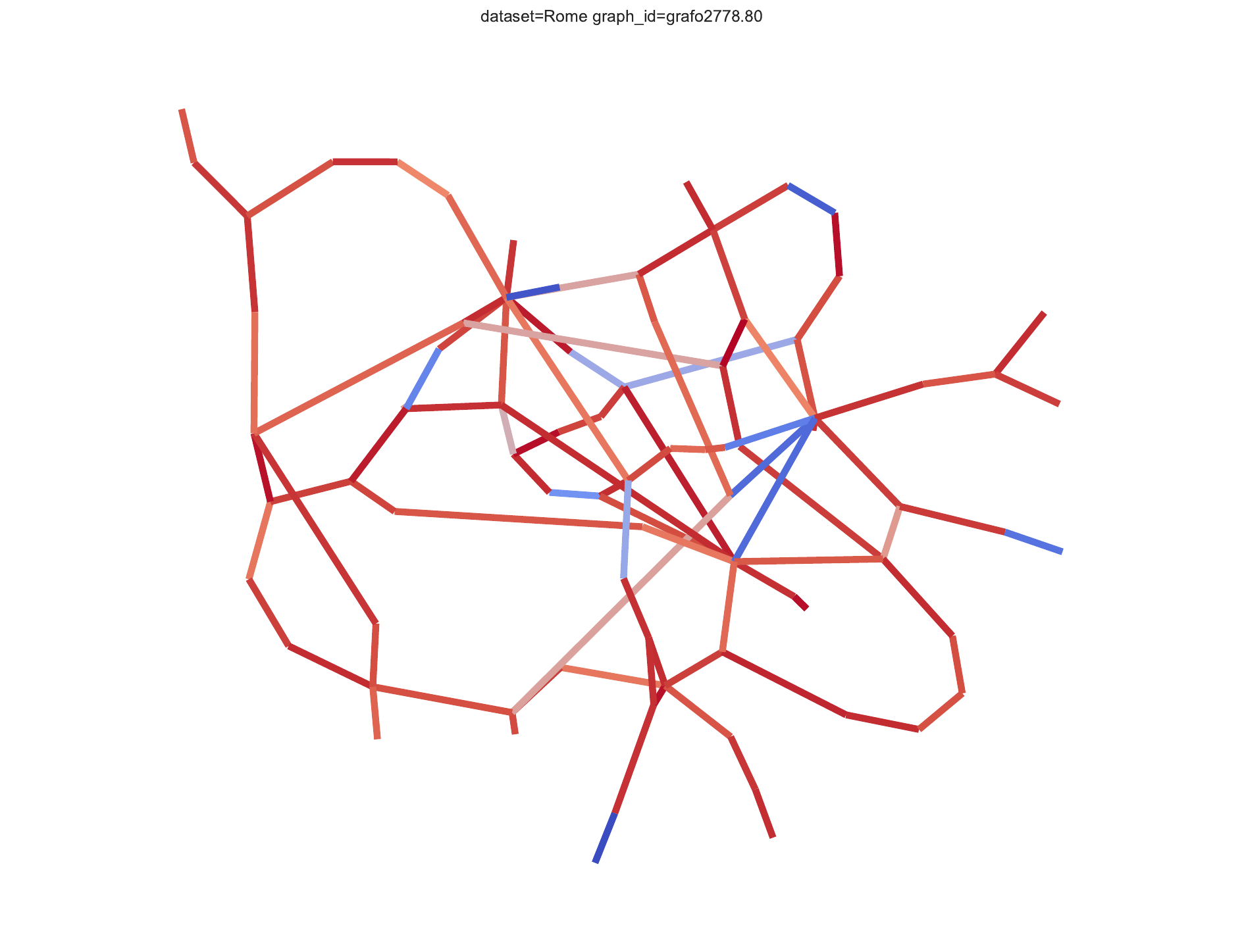} &
\imgcell{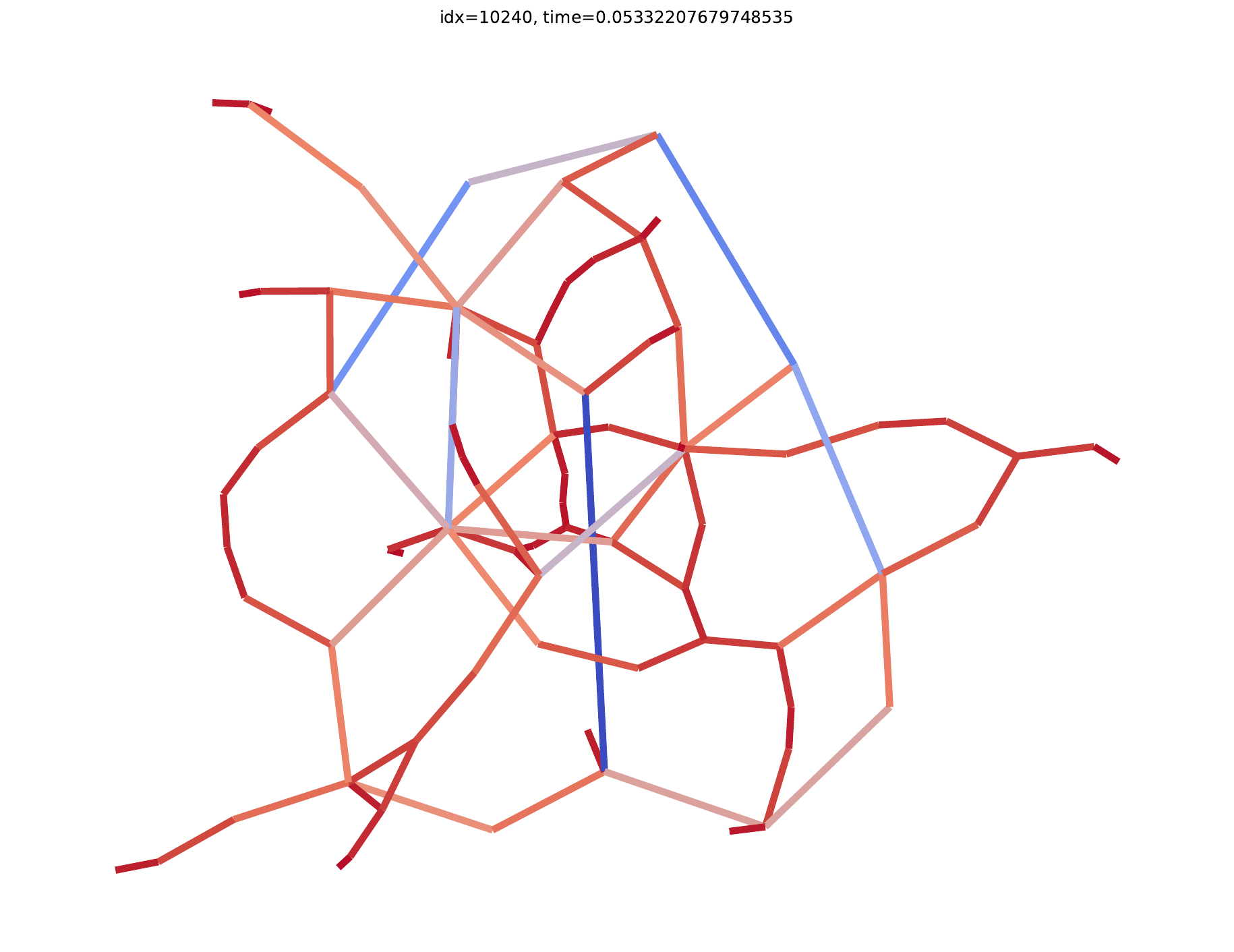} &
\imgcell{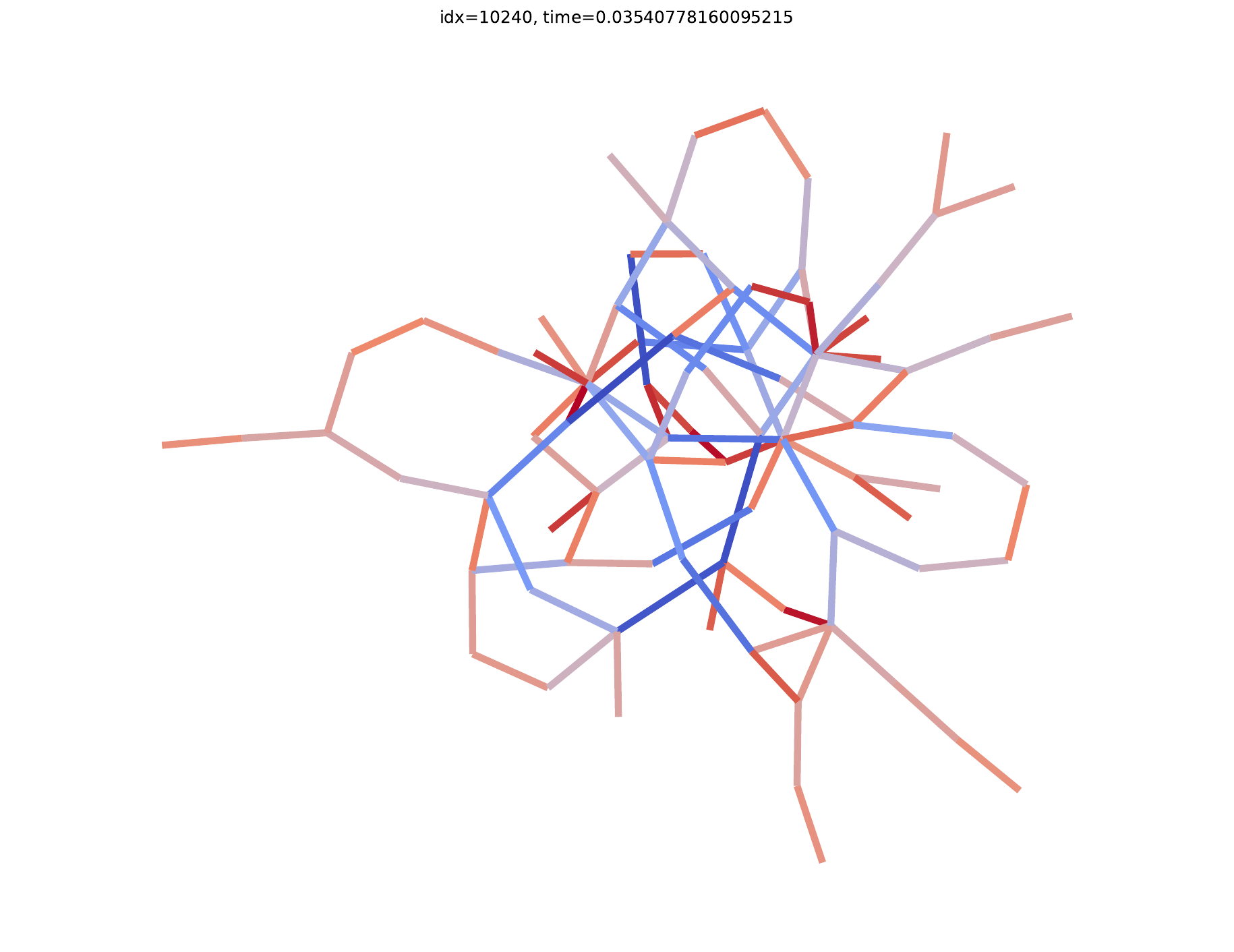} &
\imgcell{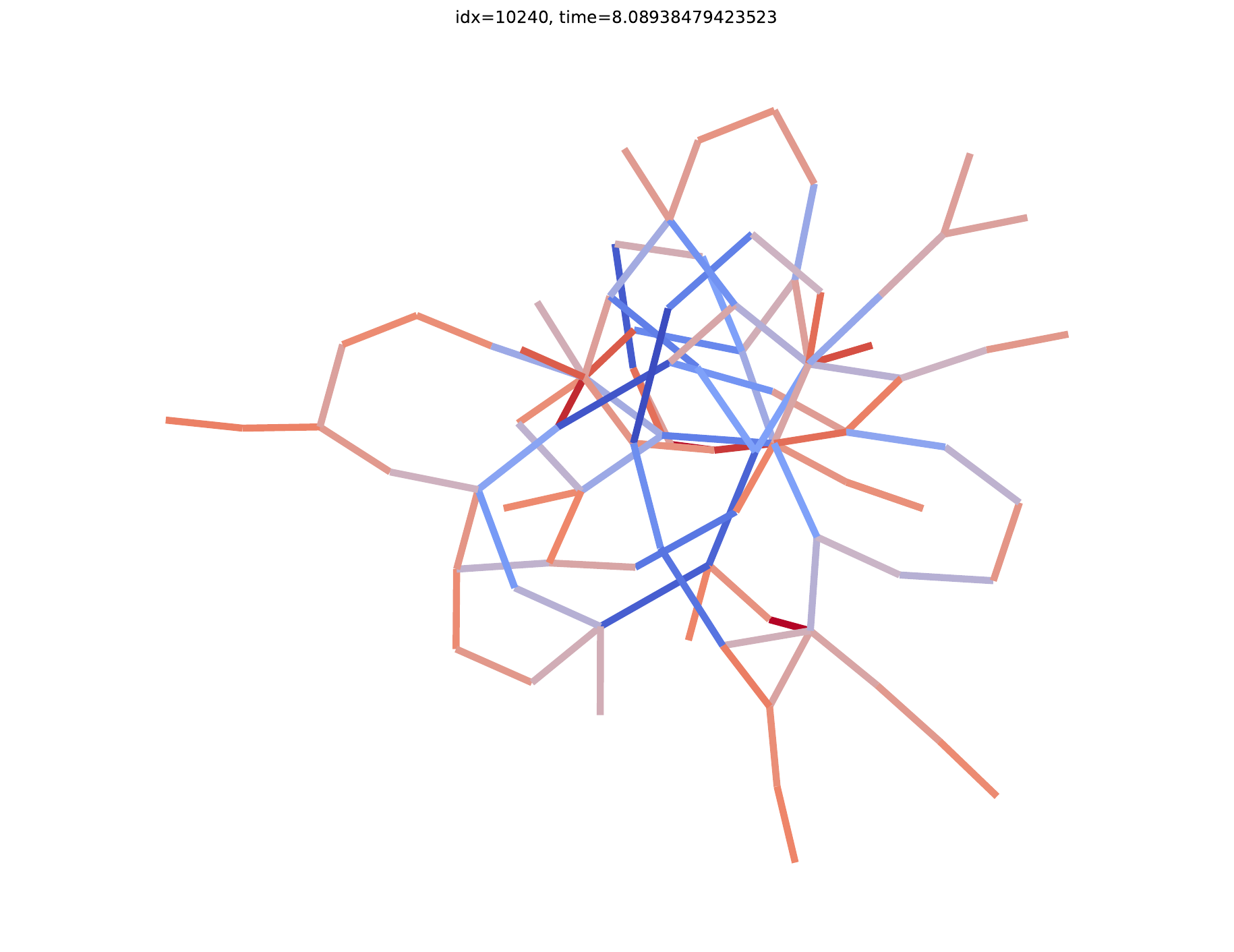} &
\imgcell{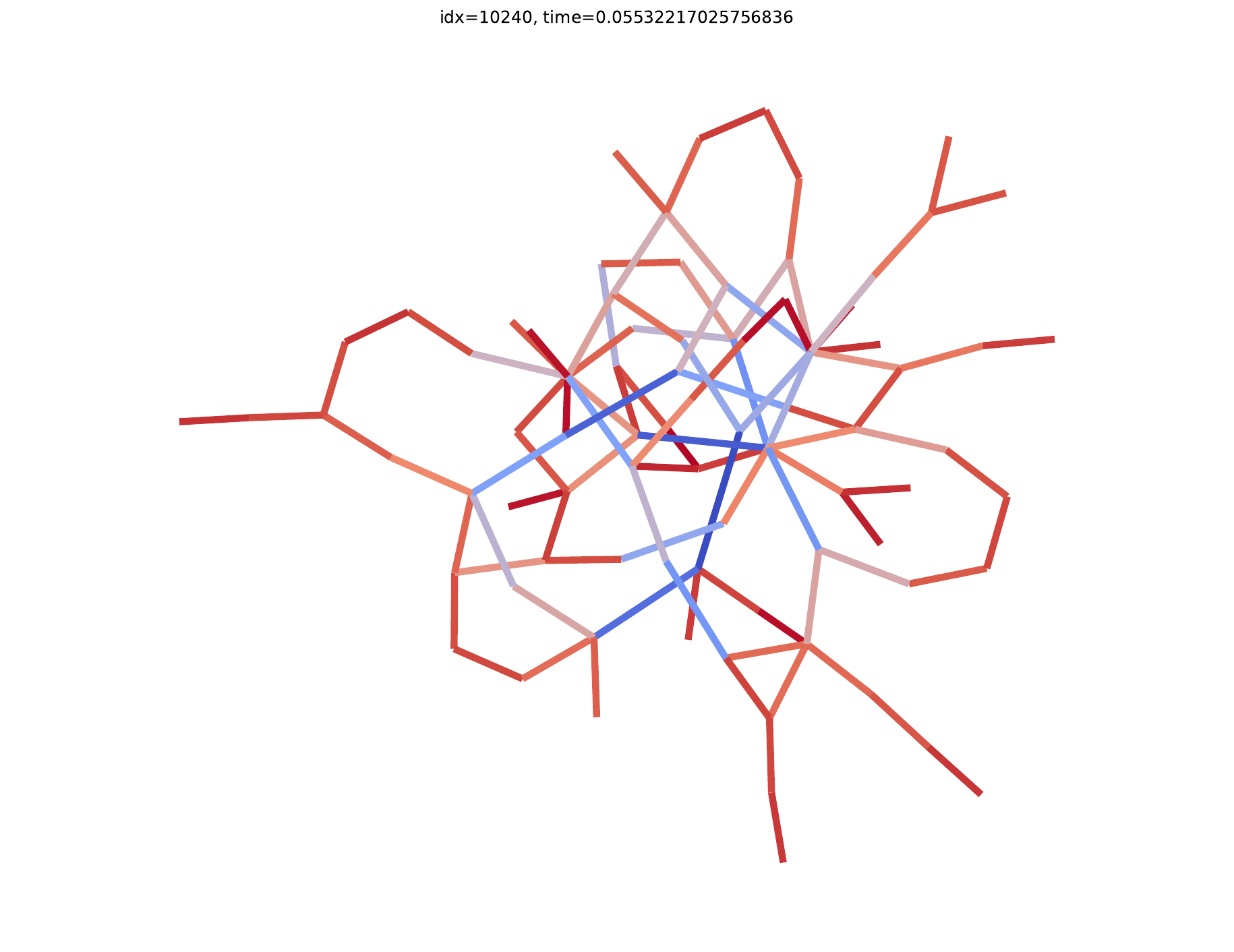} \\

&
t = 0.00s &
t = 0.59s &
t = 0.50s &
t = 0.05s &
t = 108.35s &
t = 0.04s &
t = 0.04s &
t = 0.06s &
t = 0.05s &
t = 0.04s &
t = 0.05s &
t = 0.06s \\

\makecell{\bfseries grafo5345.42\\N = 98\\M = 136} &
\imgcell{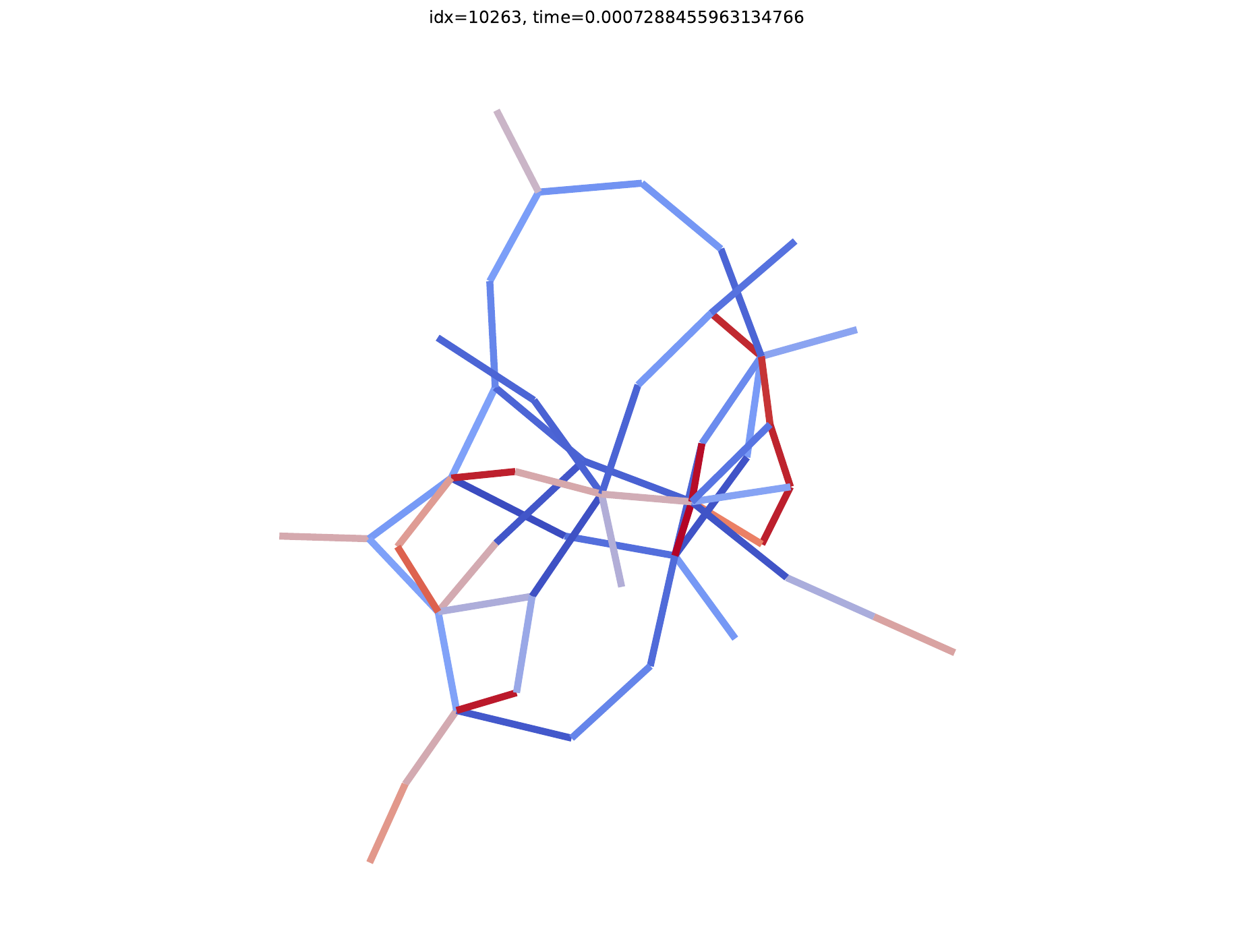} &
\imgcell{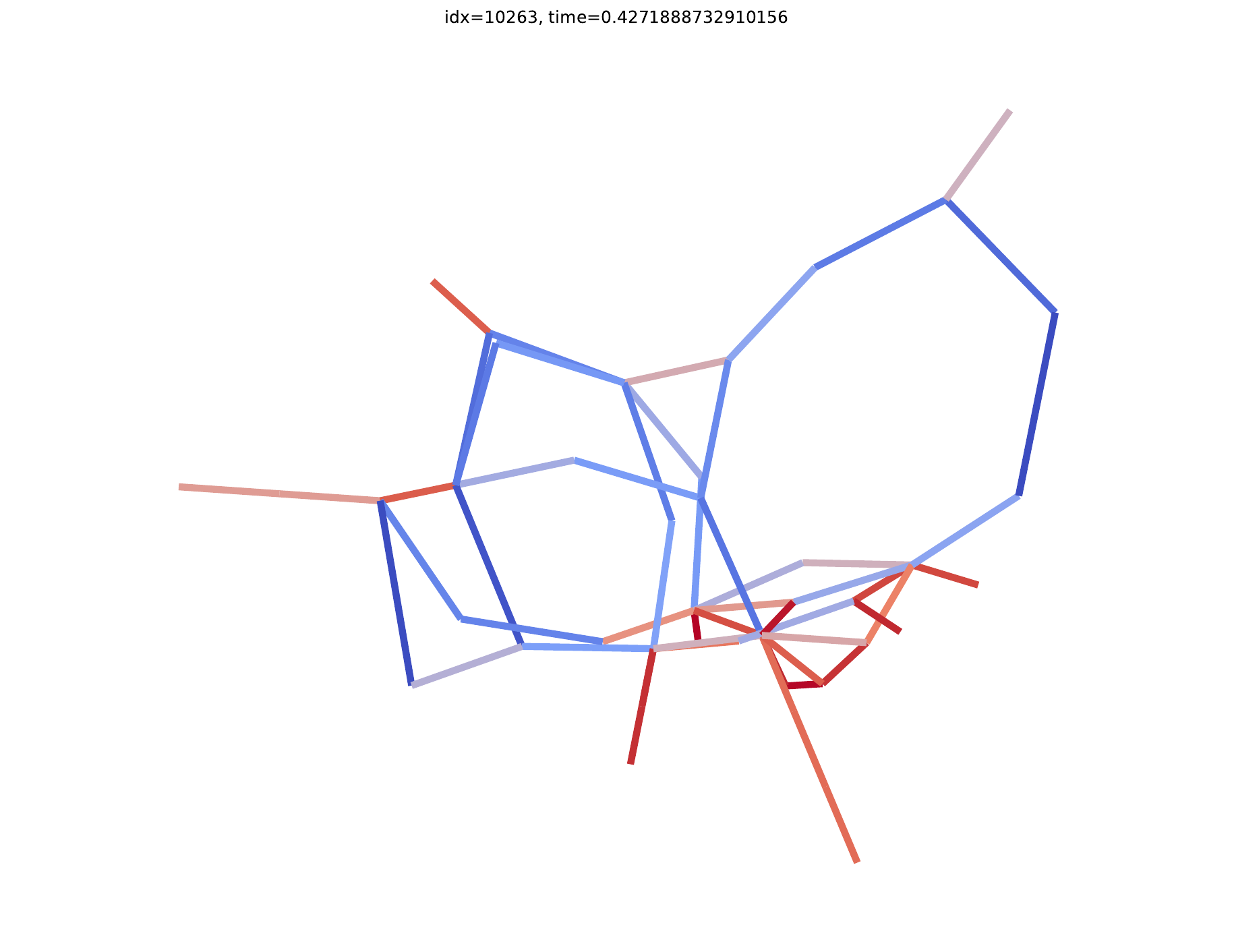} &
\imgcell{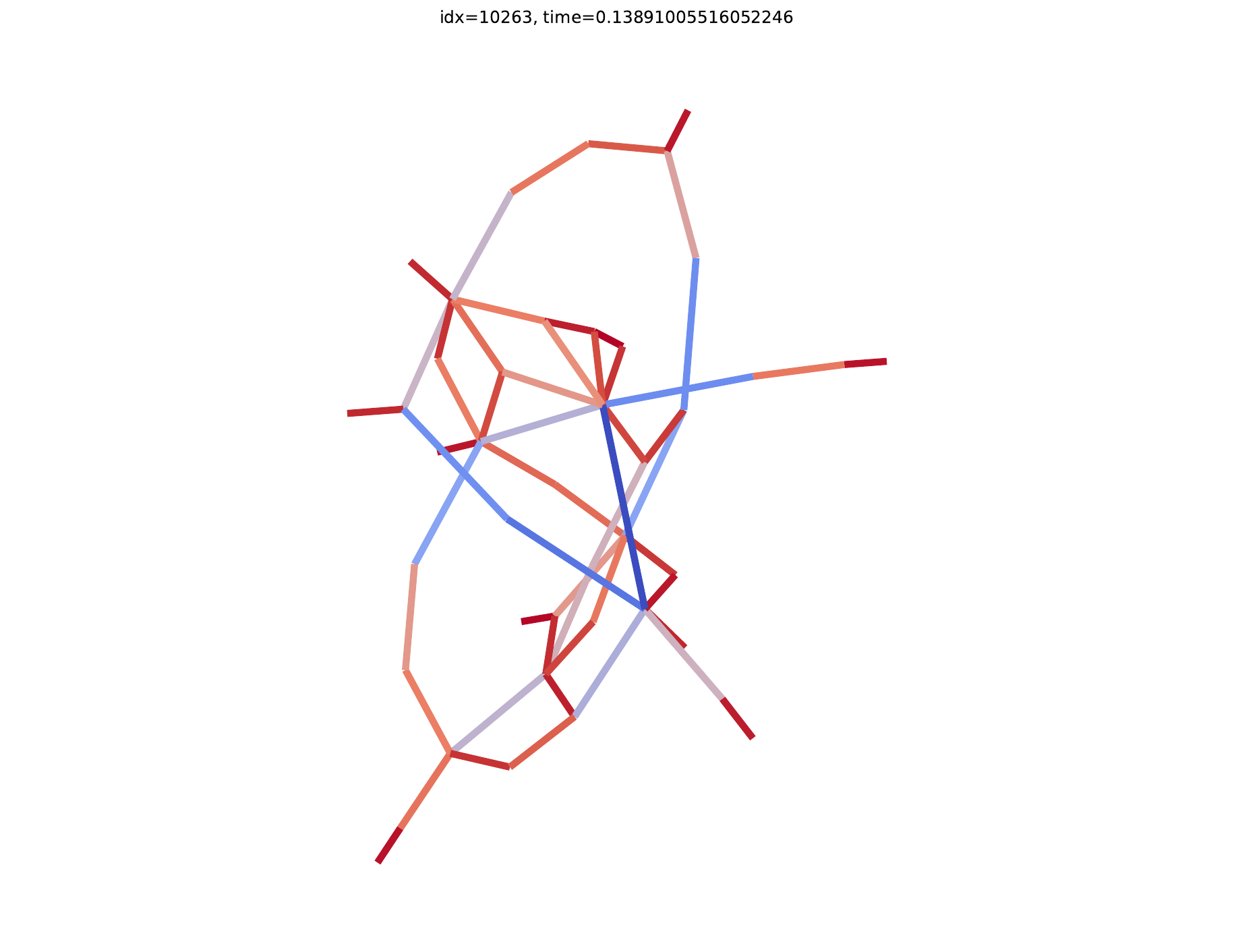} &
\imgcell{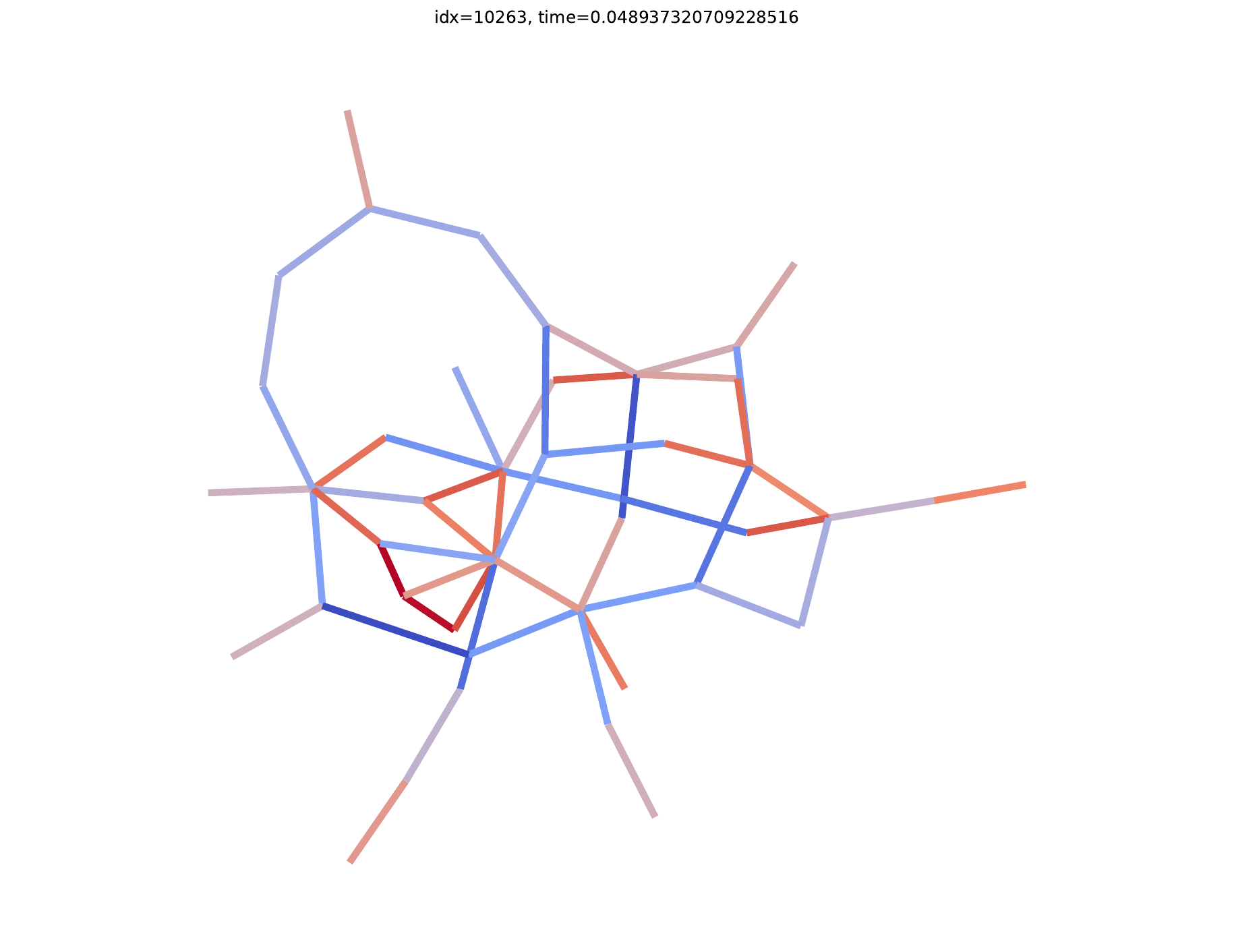} &
\imgcell{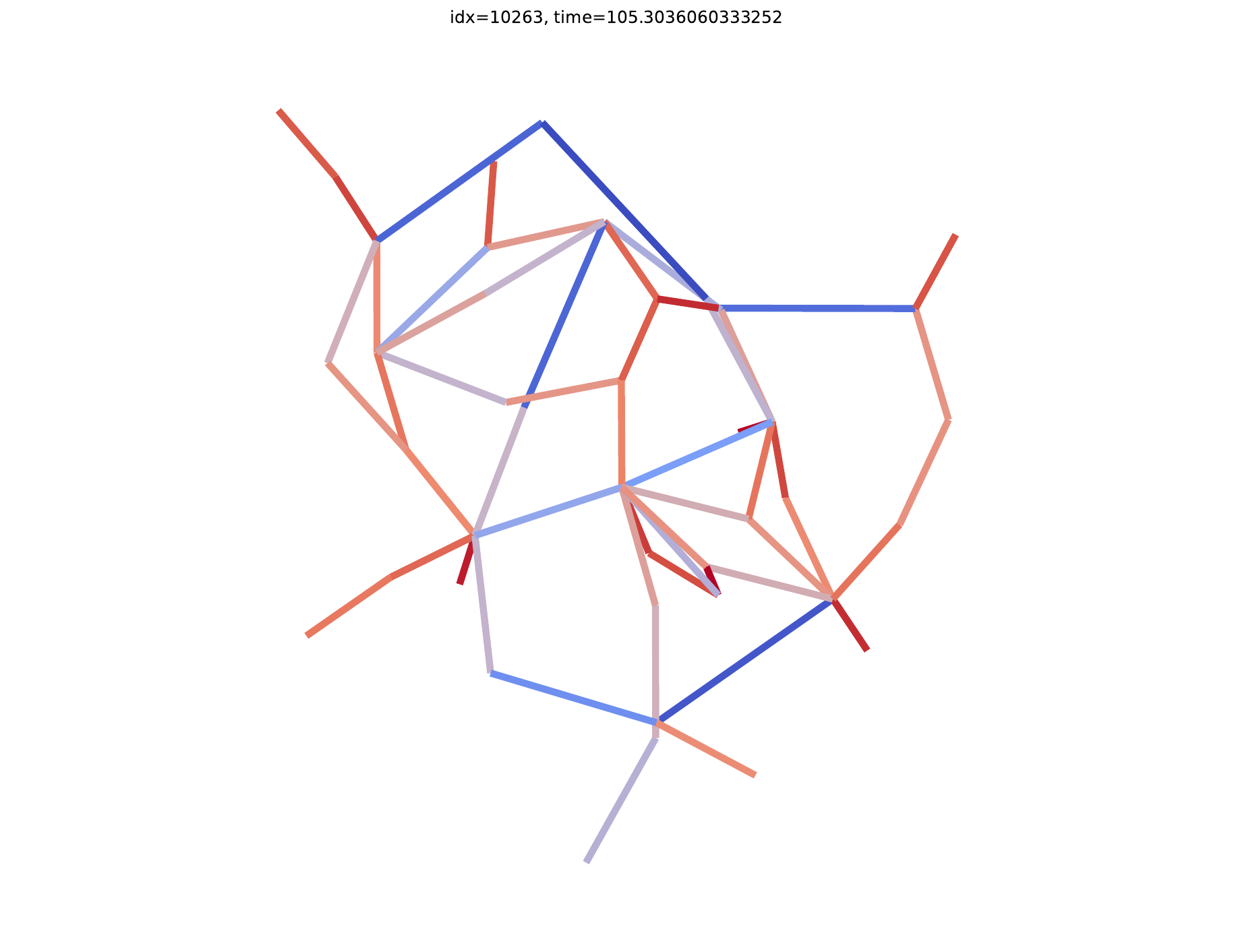} &
\imgcell{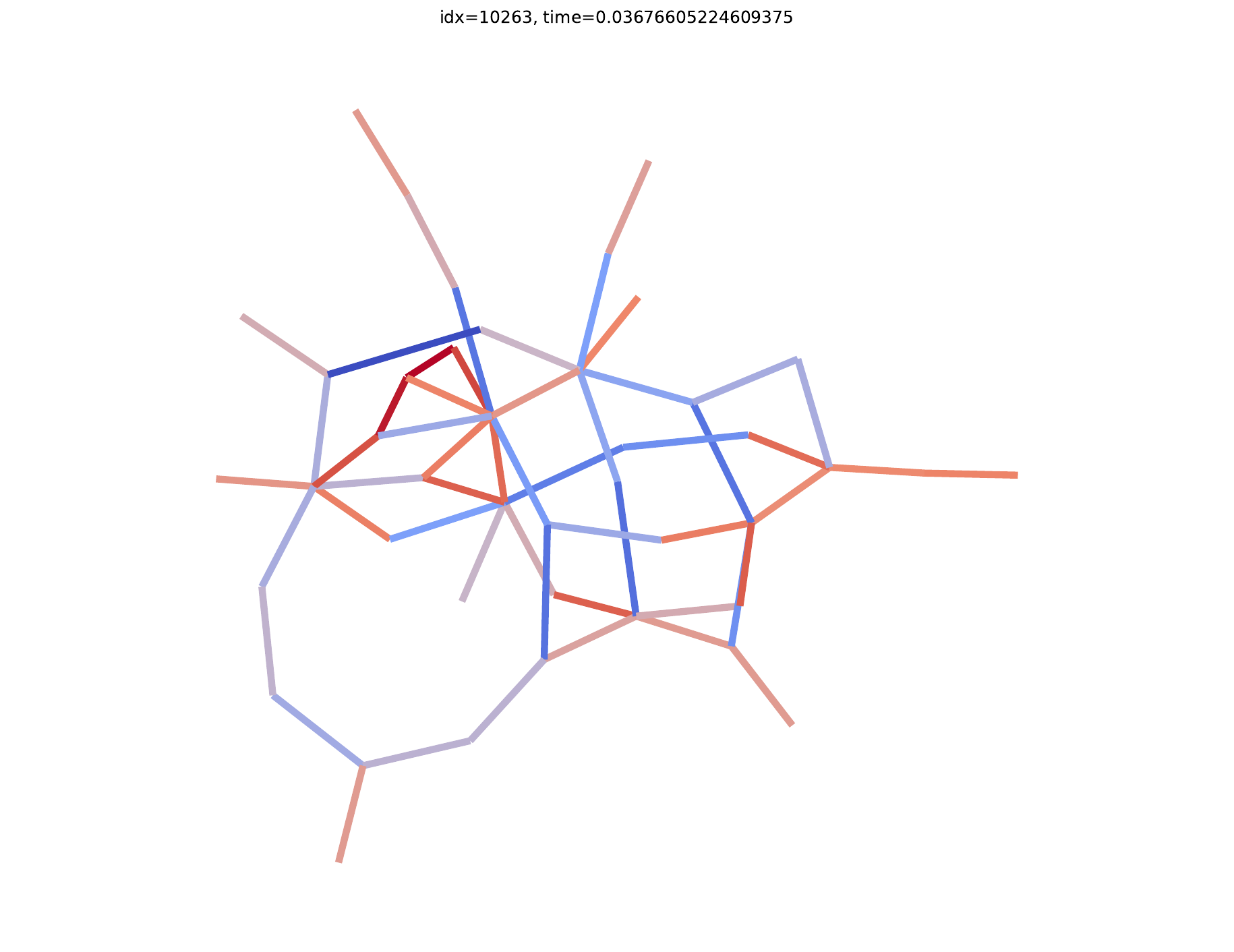} &
\imgcell{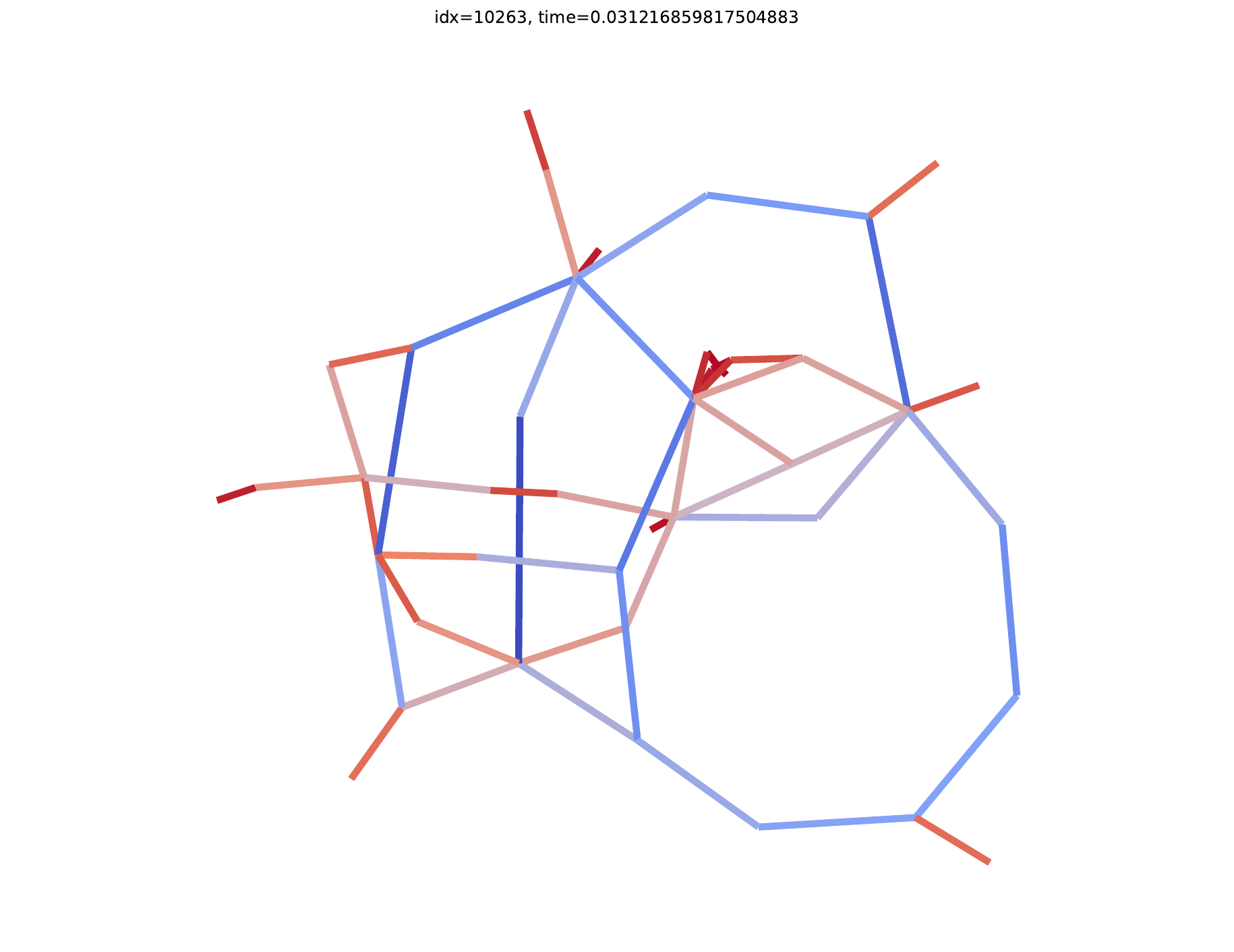} &
\imgcell{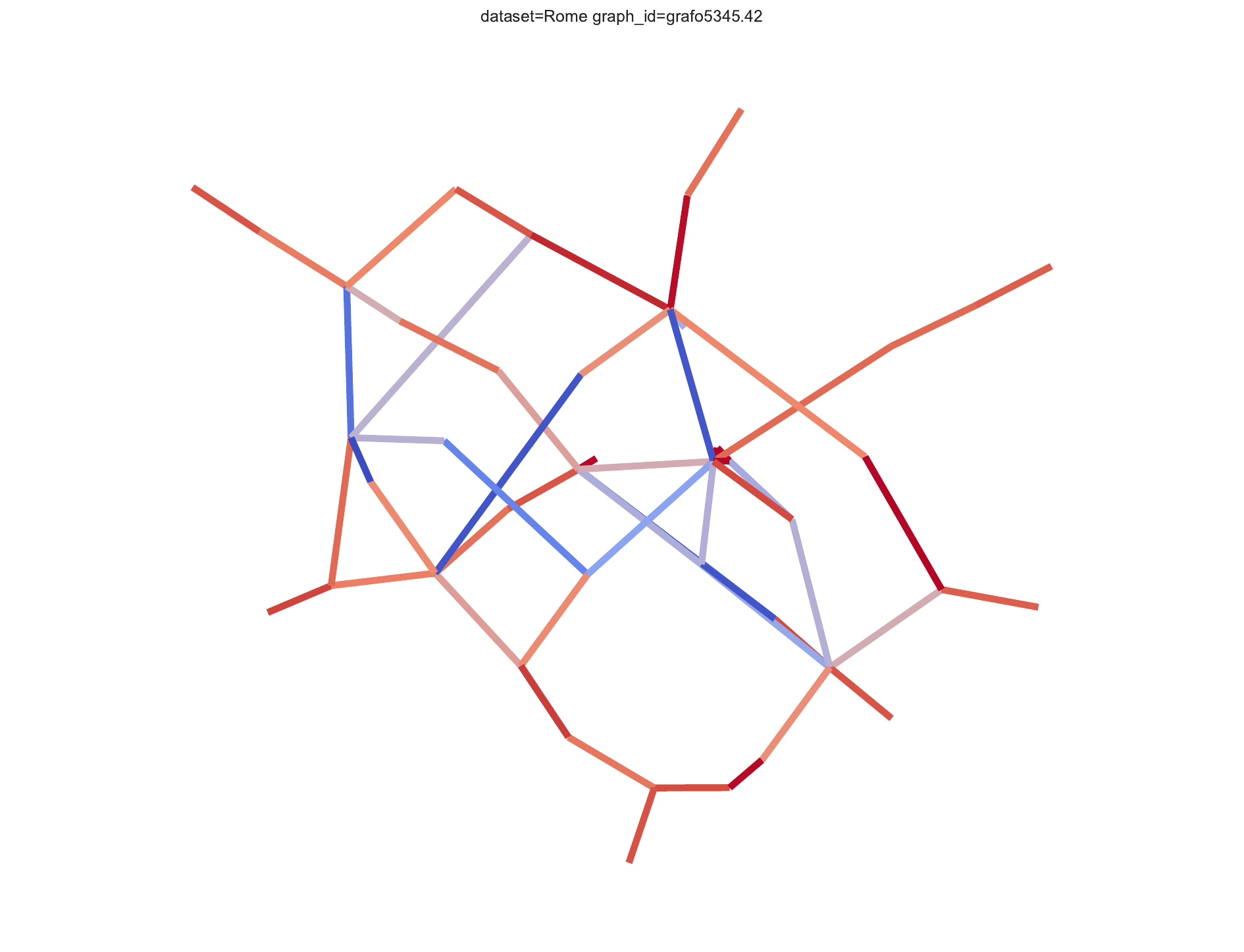} &
\imgcell{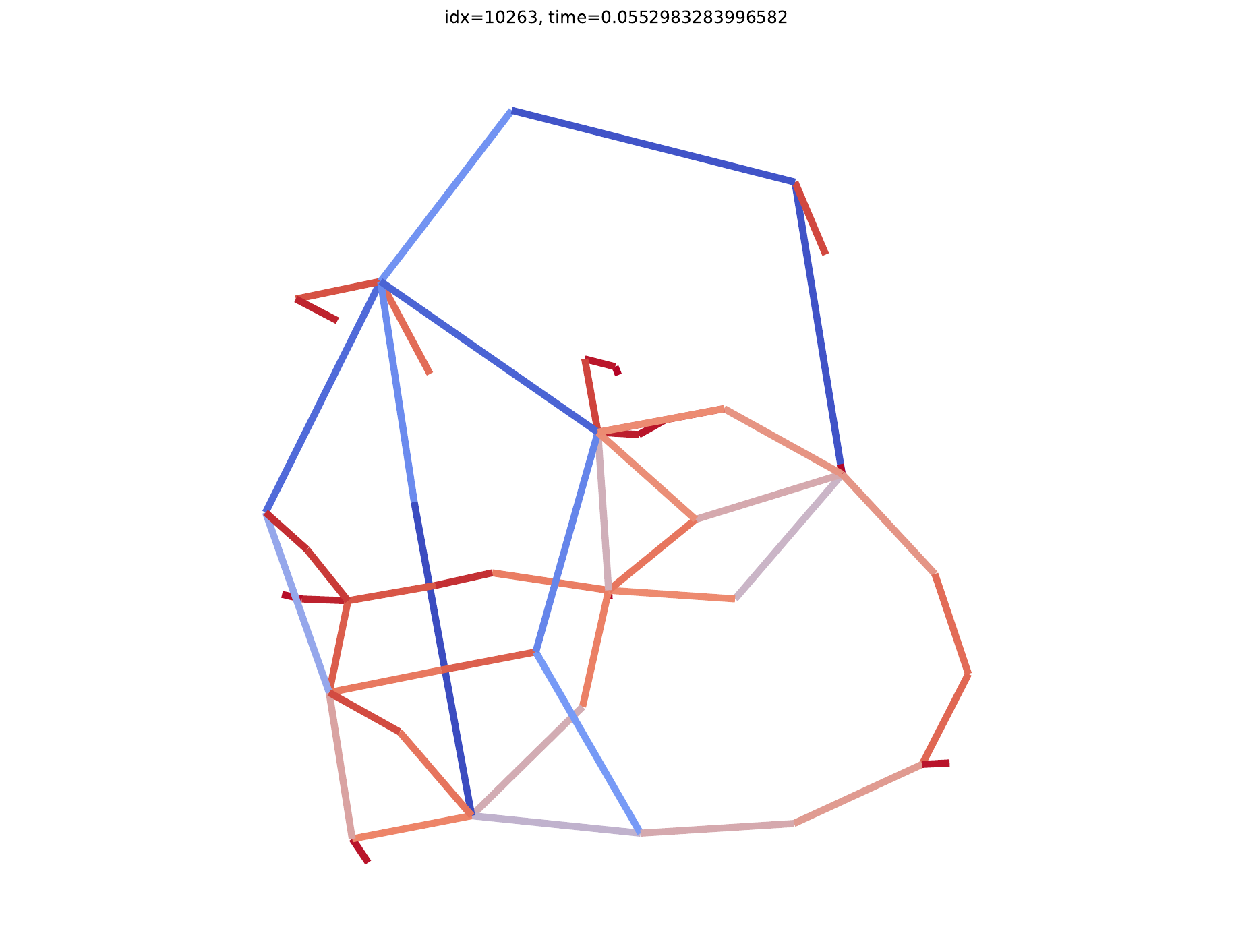} &
\imgcell{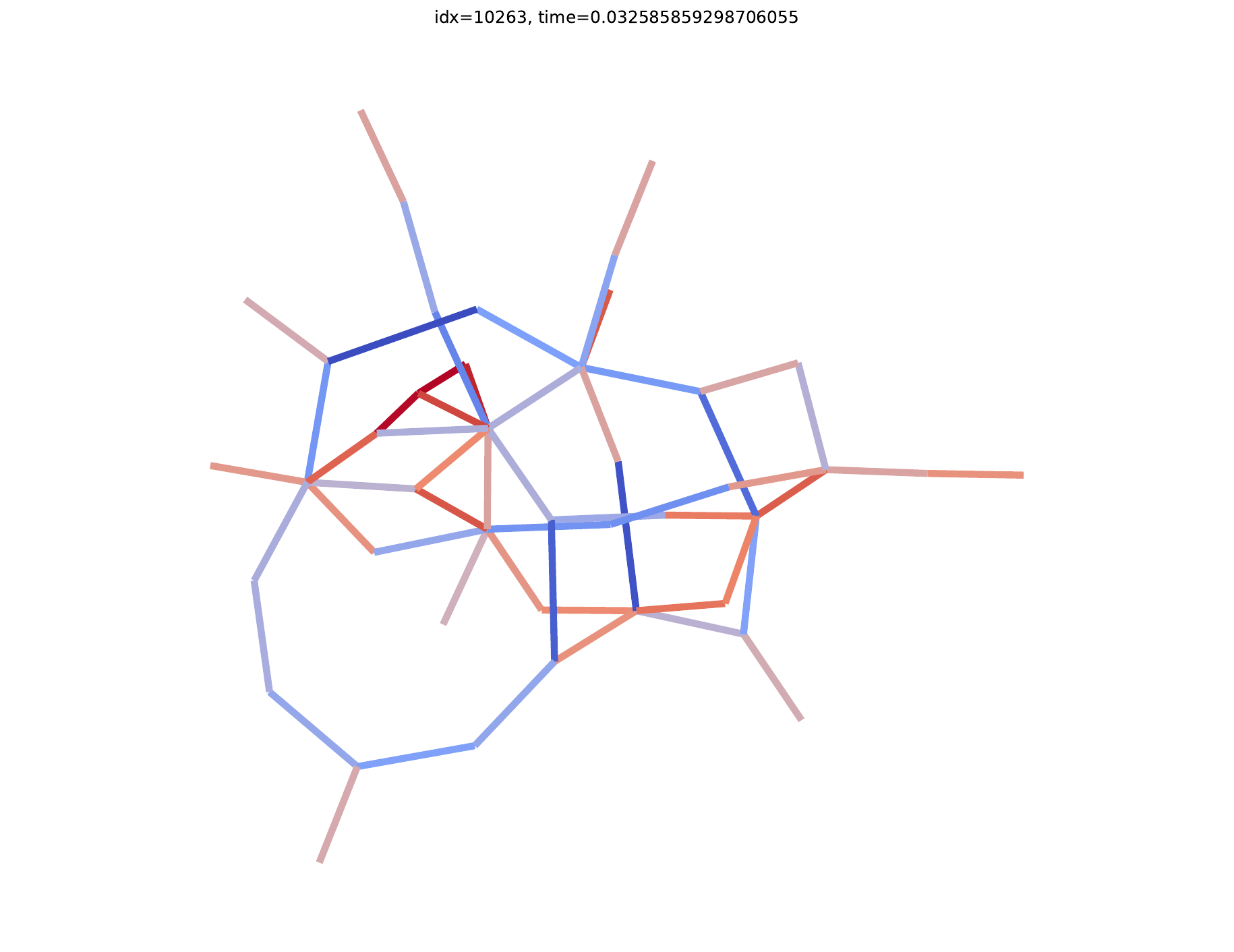} &
\imgcell{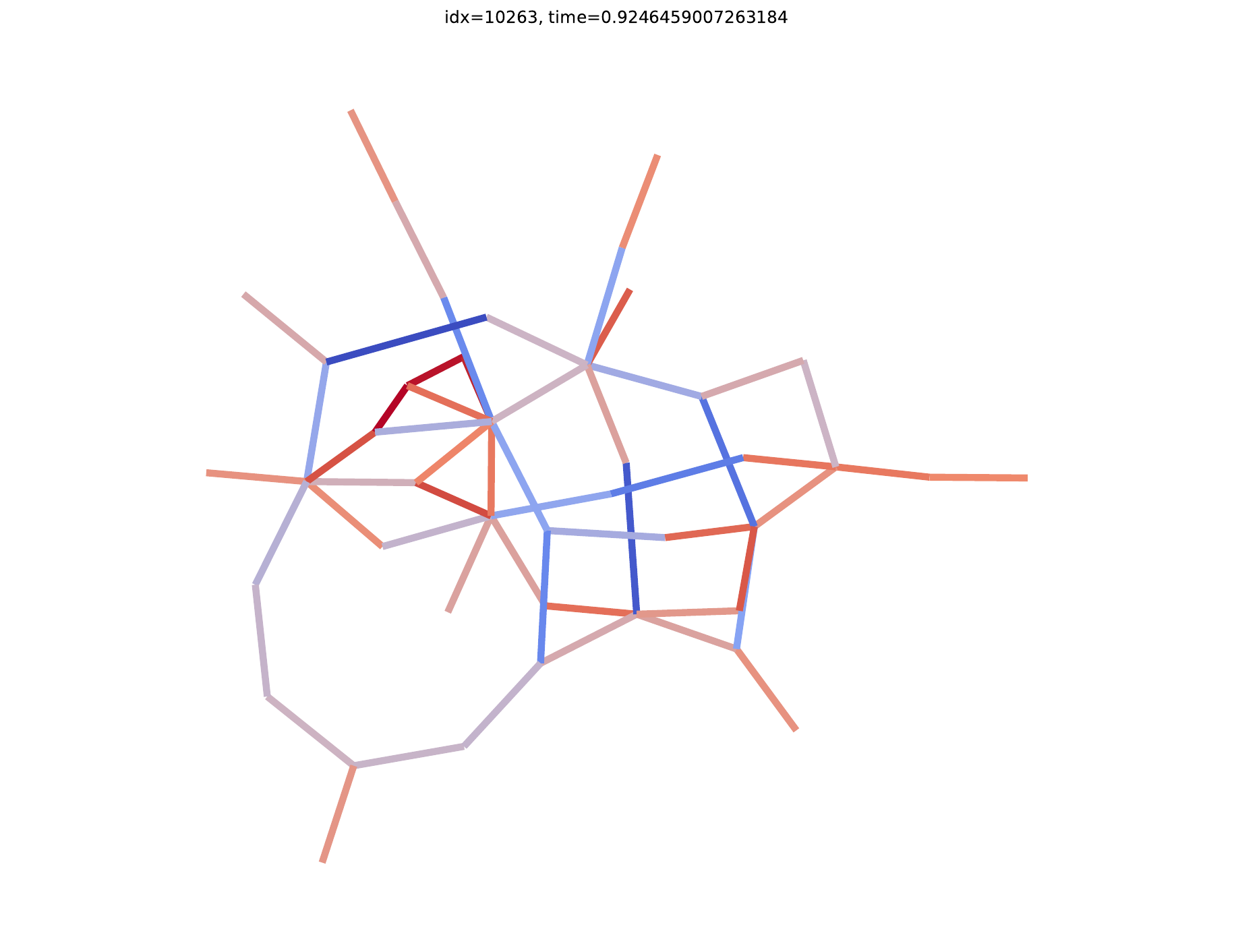} &
\imgcell{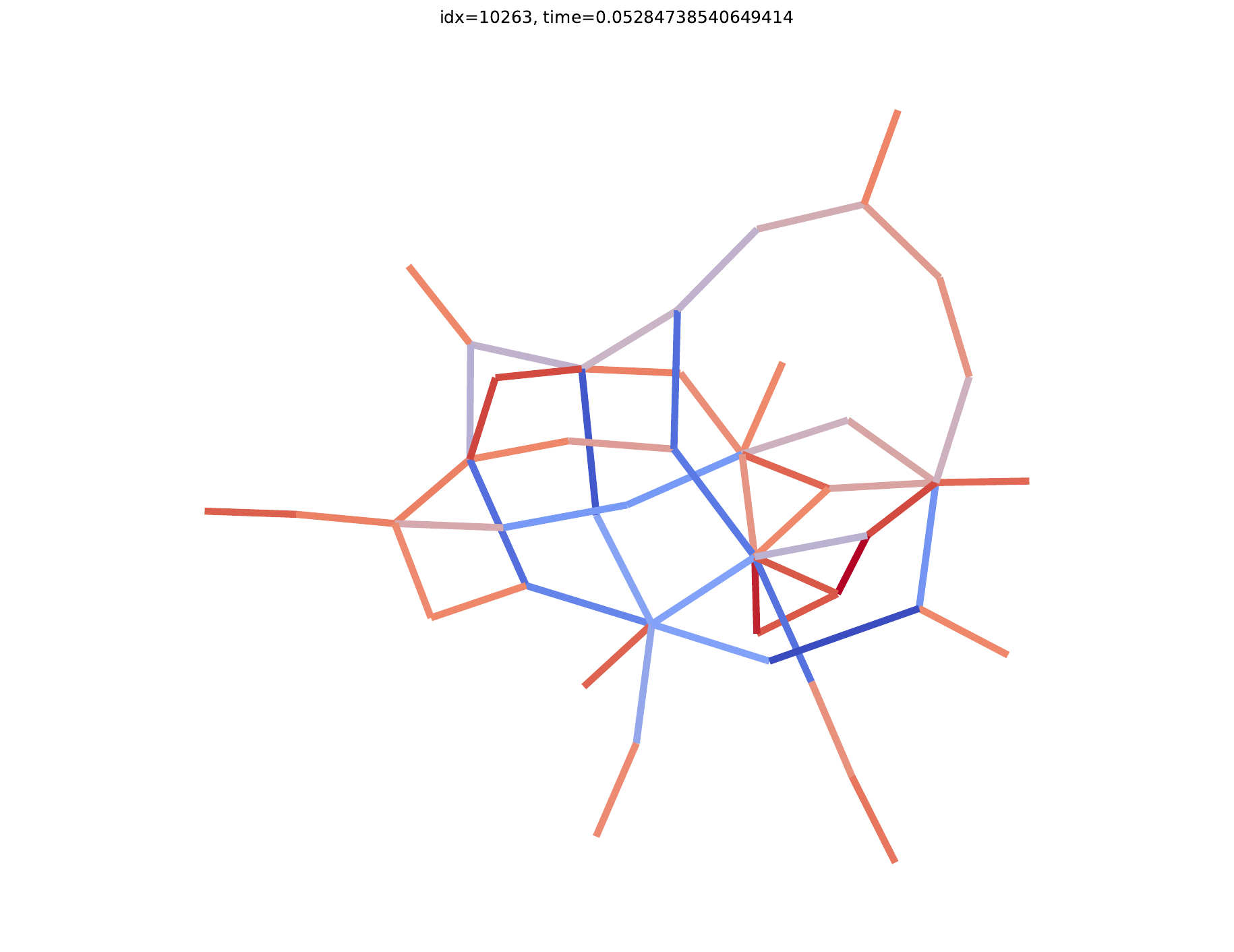} \\

&
t = 0.00s &
t = 0.43s &
t = 0.14s &
t = 0.05s &
t = 105.30s &
t = 0.04s &
t = 0.03s &
t = 0.05s &
t = 0.06s &
t = 0.03s &
t = 0.03s &
t = 0.05s \\

\makecell{\bfseries grafo11611.32\\N = 95\\M = 129} &
\imgcell{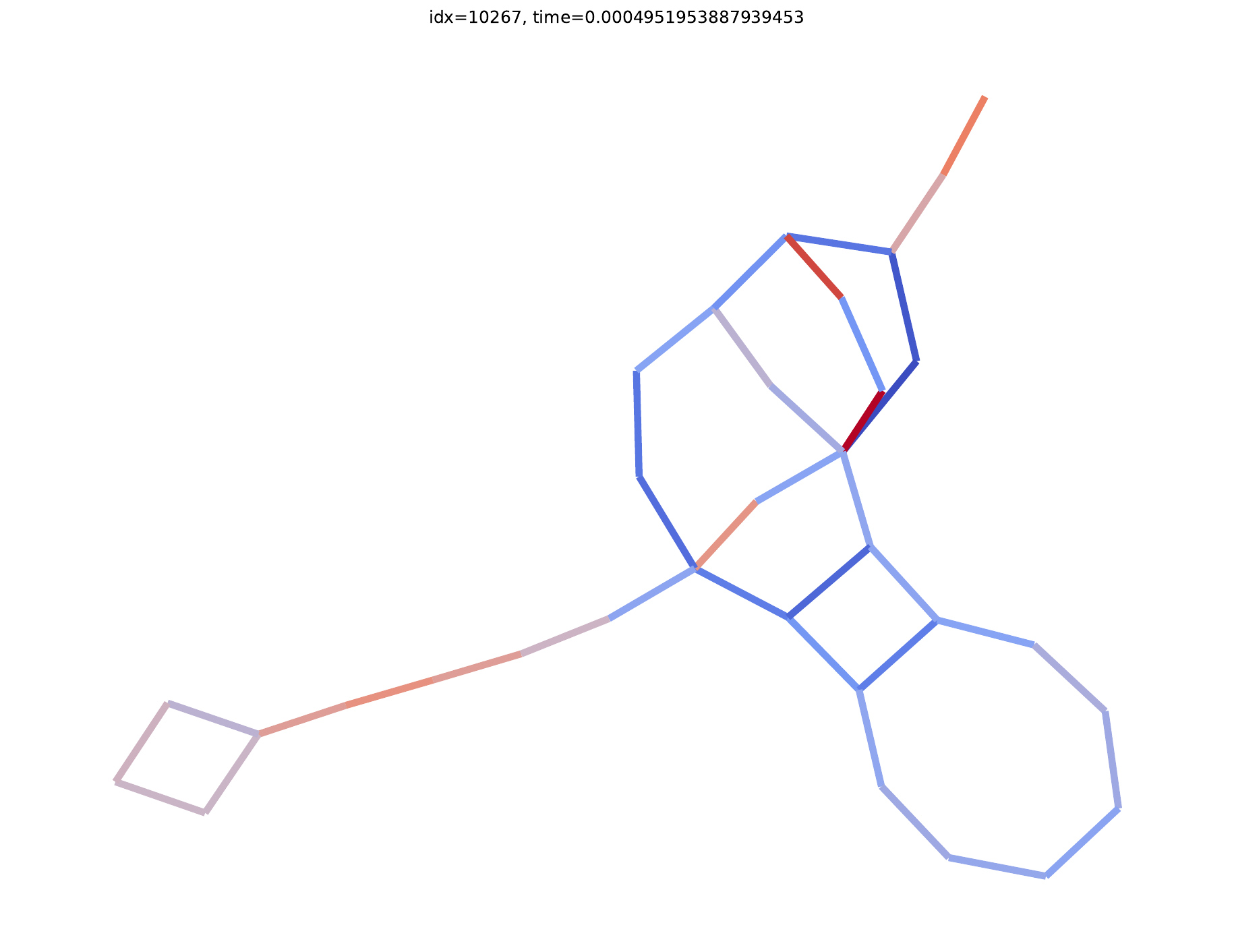} &
\imgcell{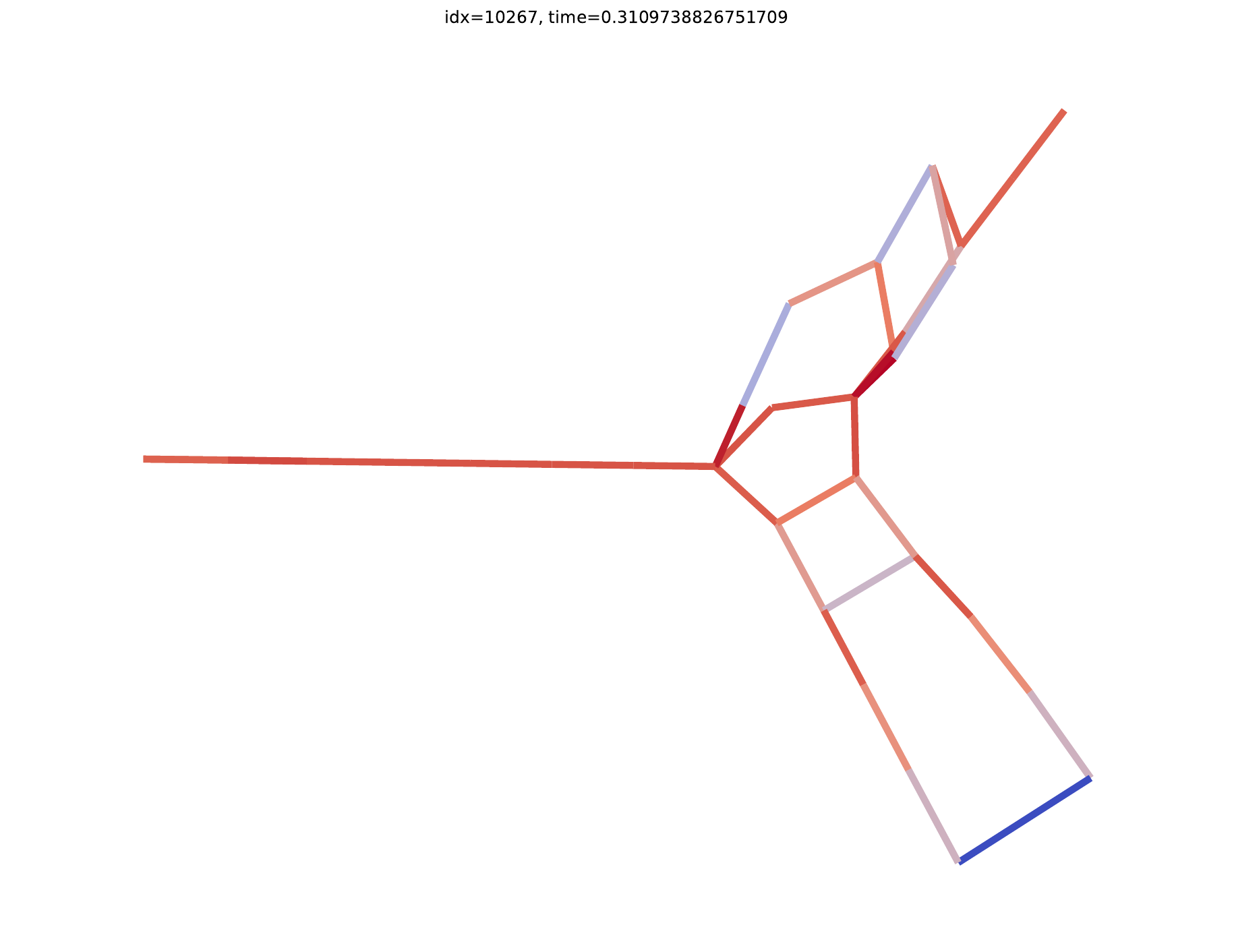} &
\imgcell{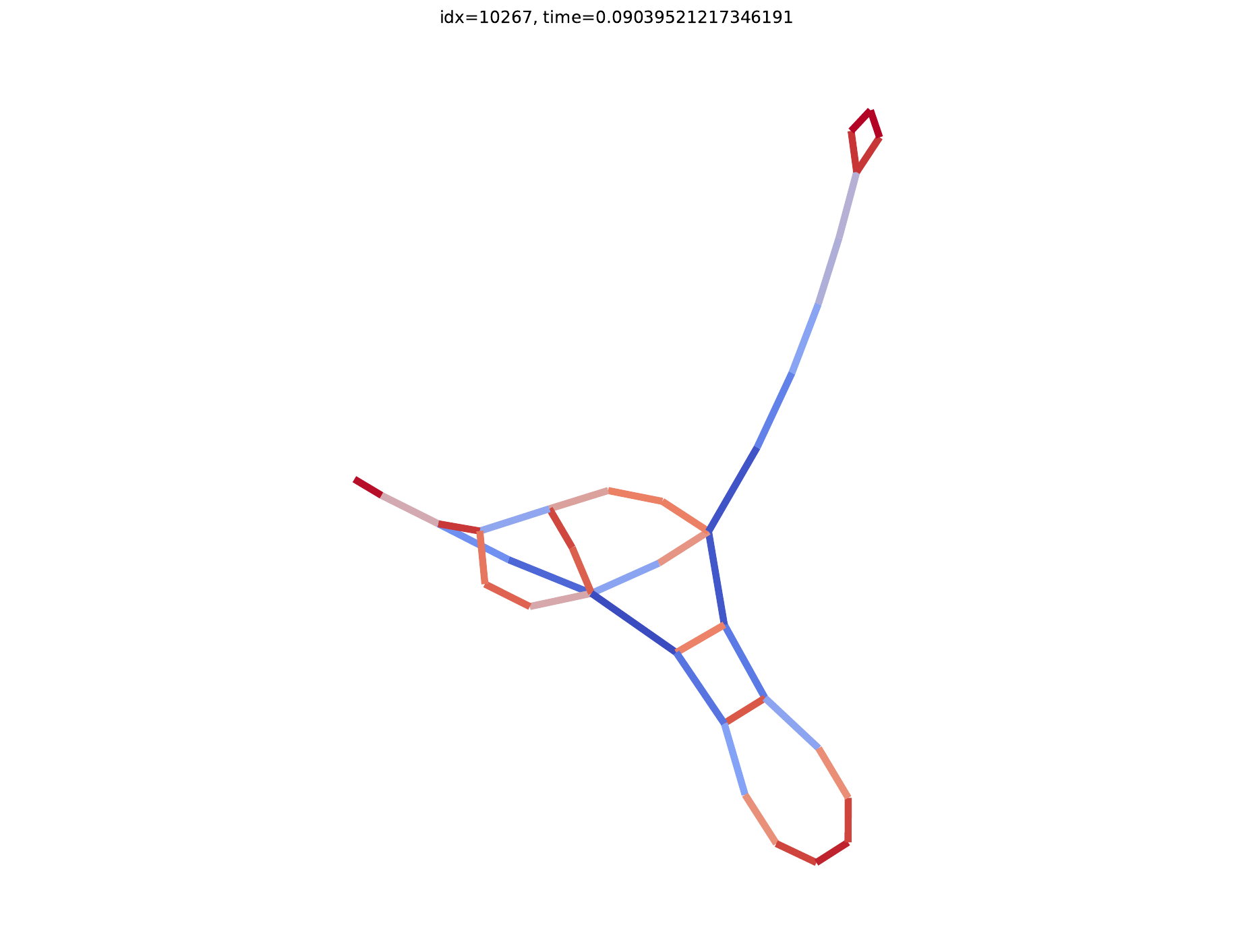} &
\imgcell{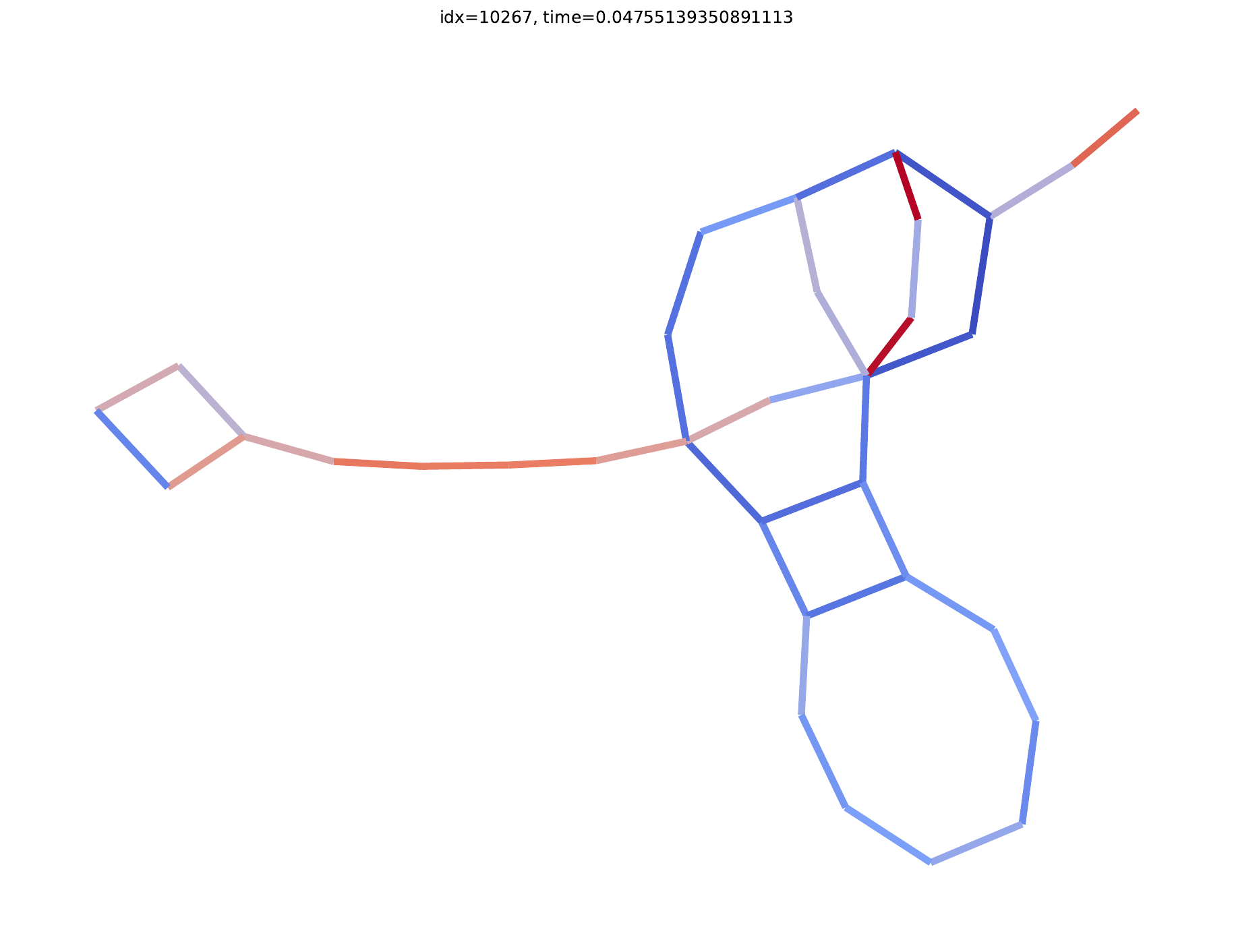} &
\imgcell{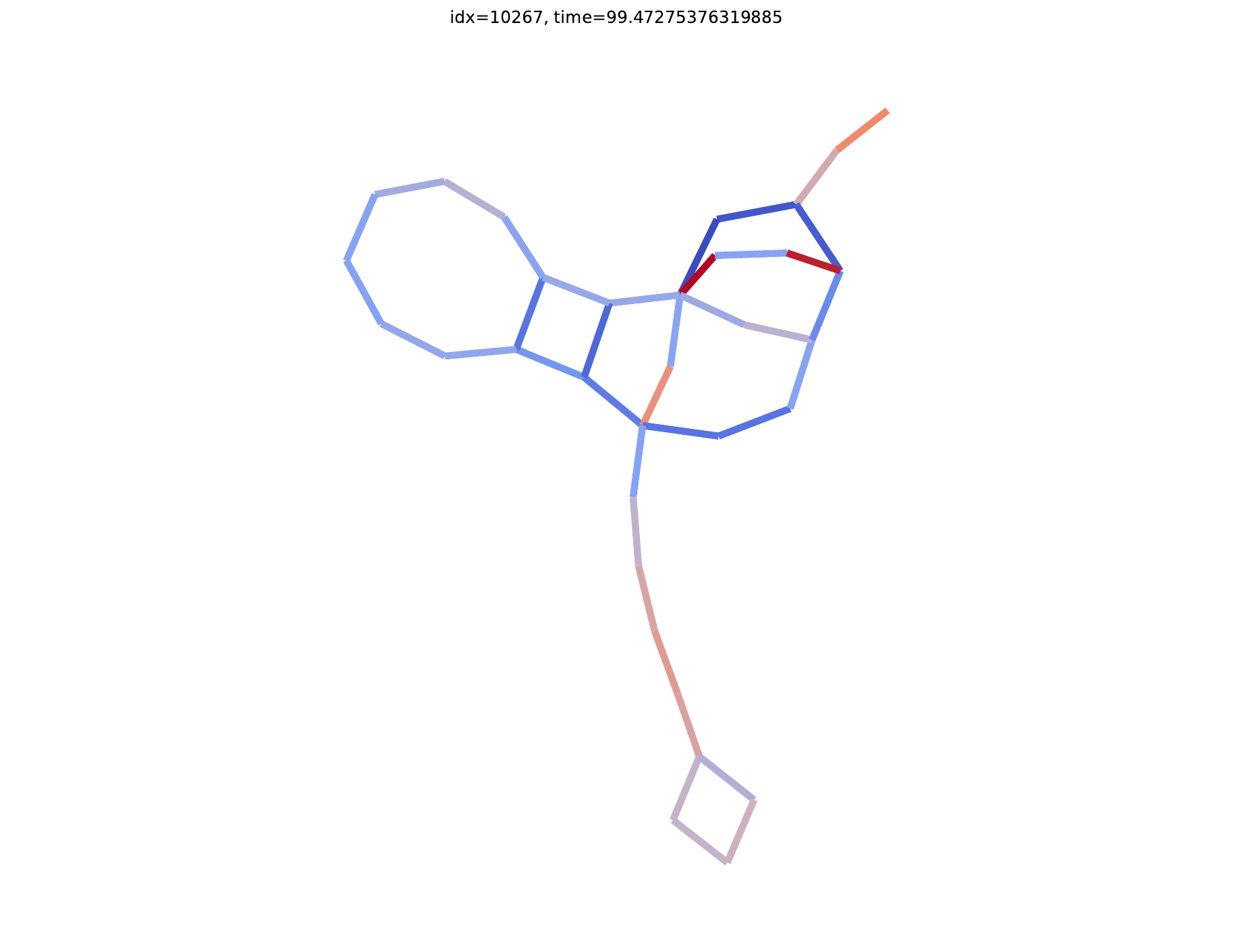} &
\imgcell{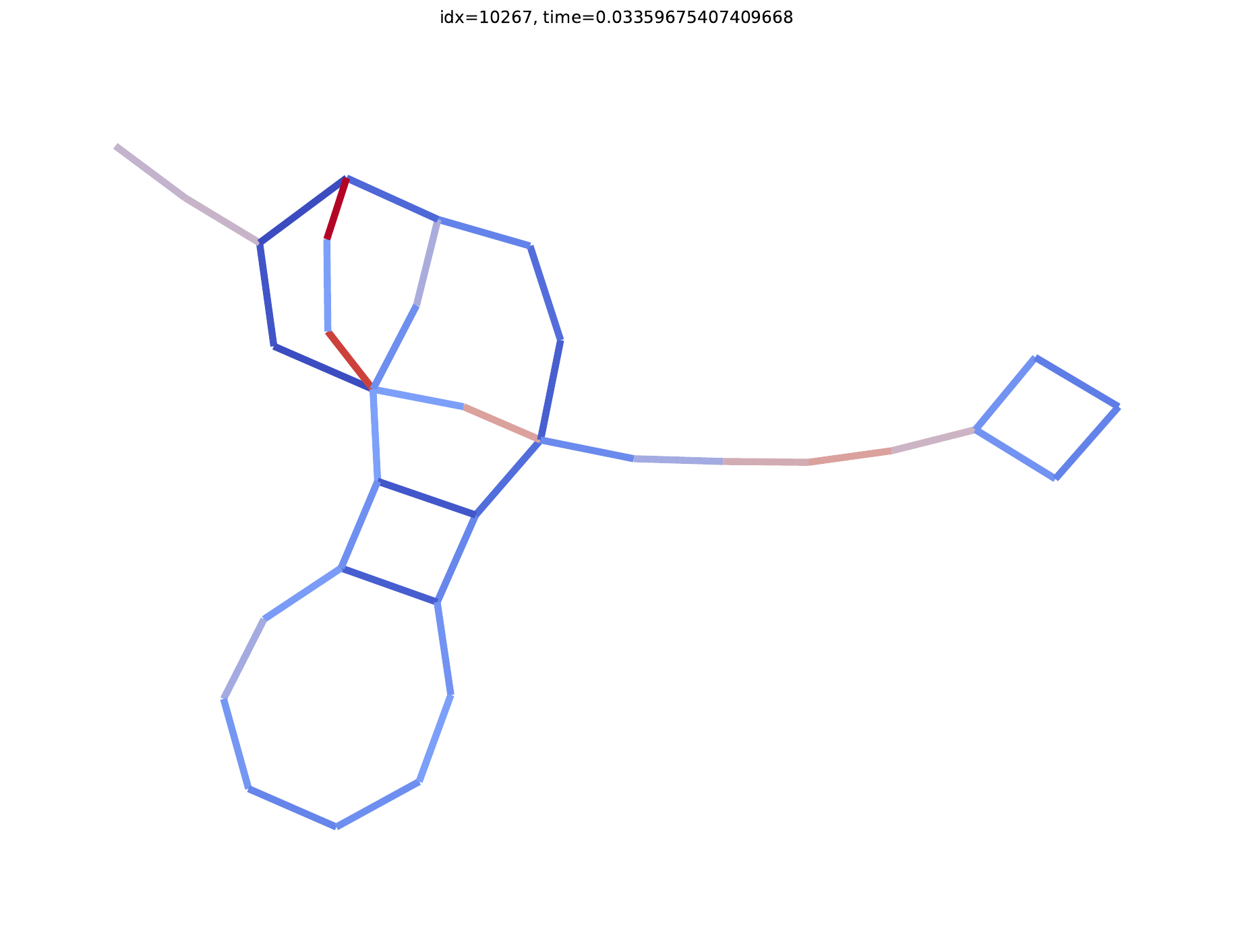} &
\imgcell{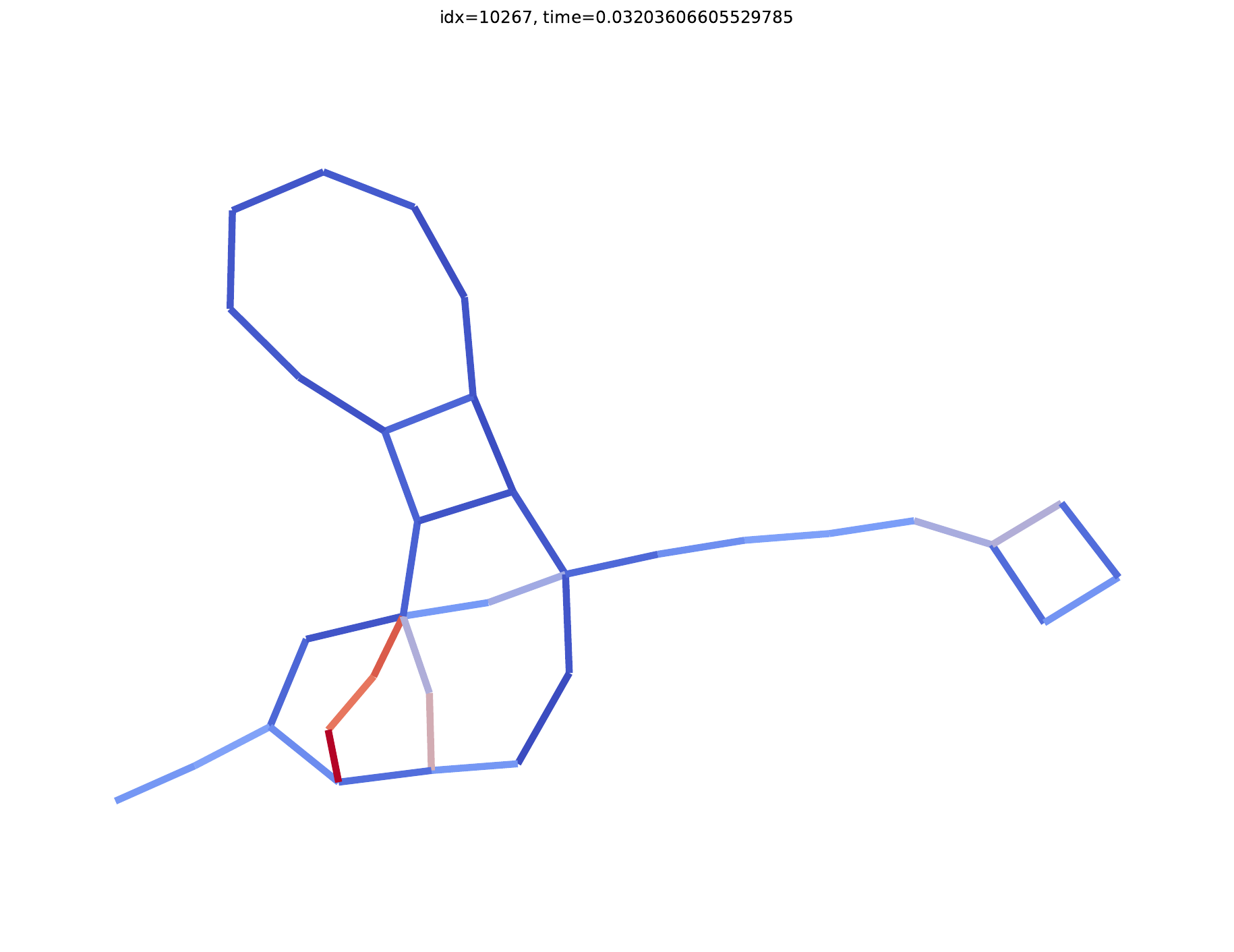} &
\imgcell{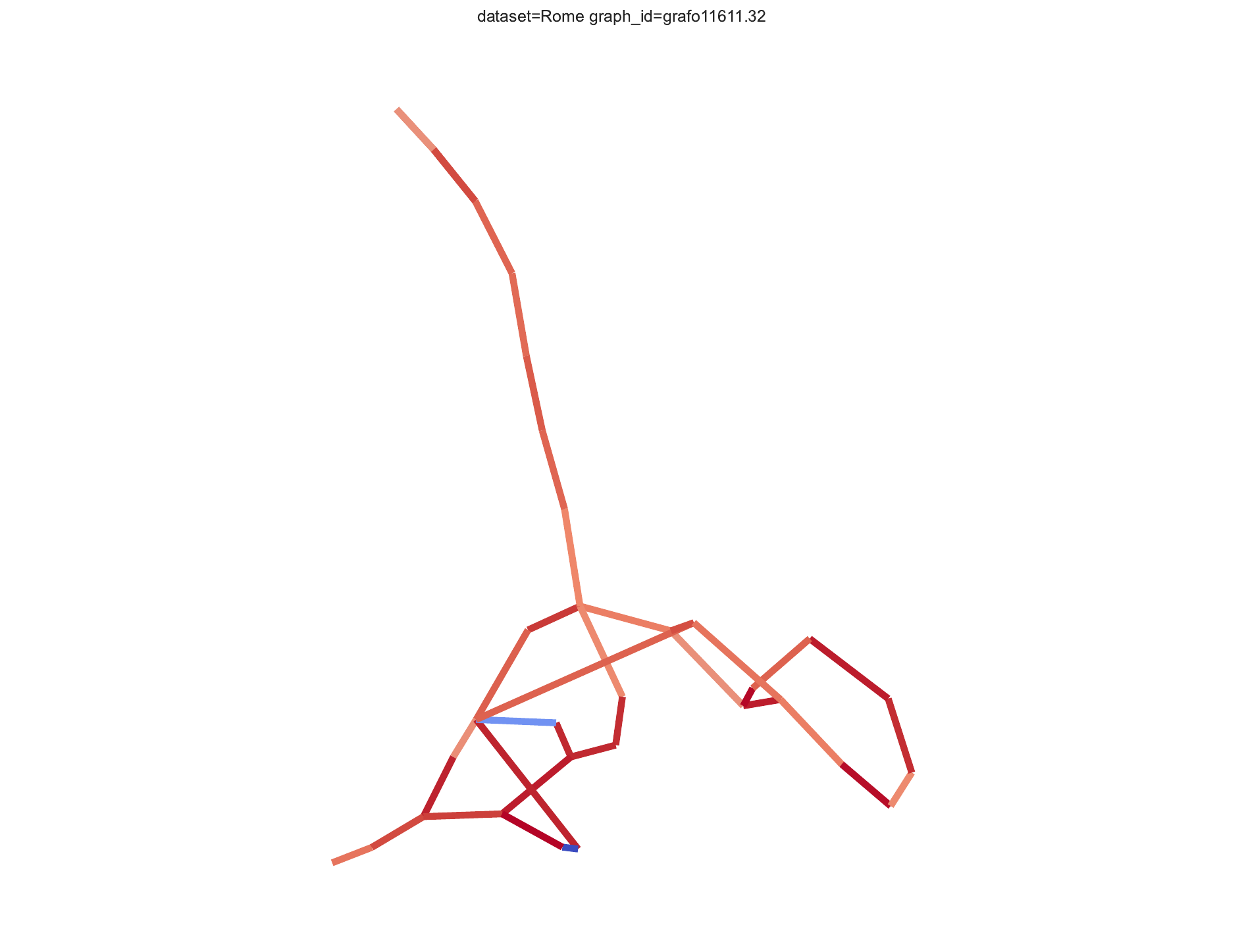} &
\imgcell{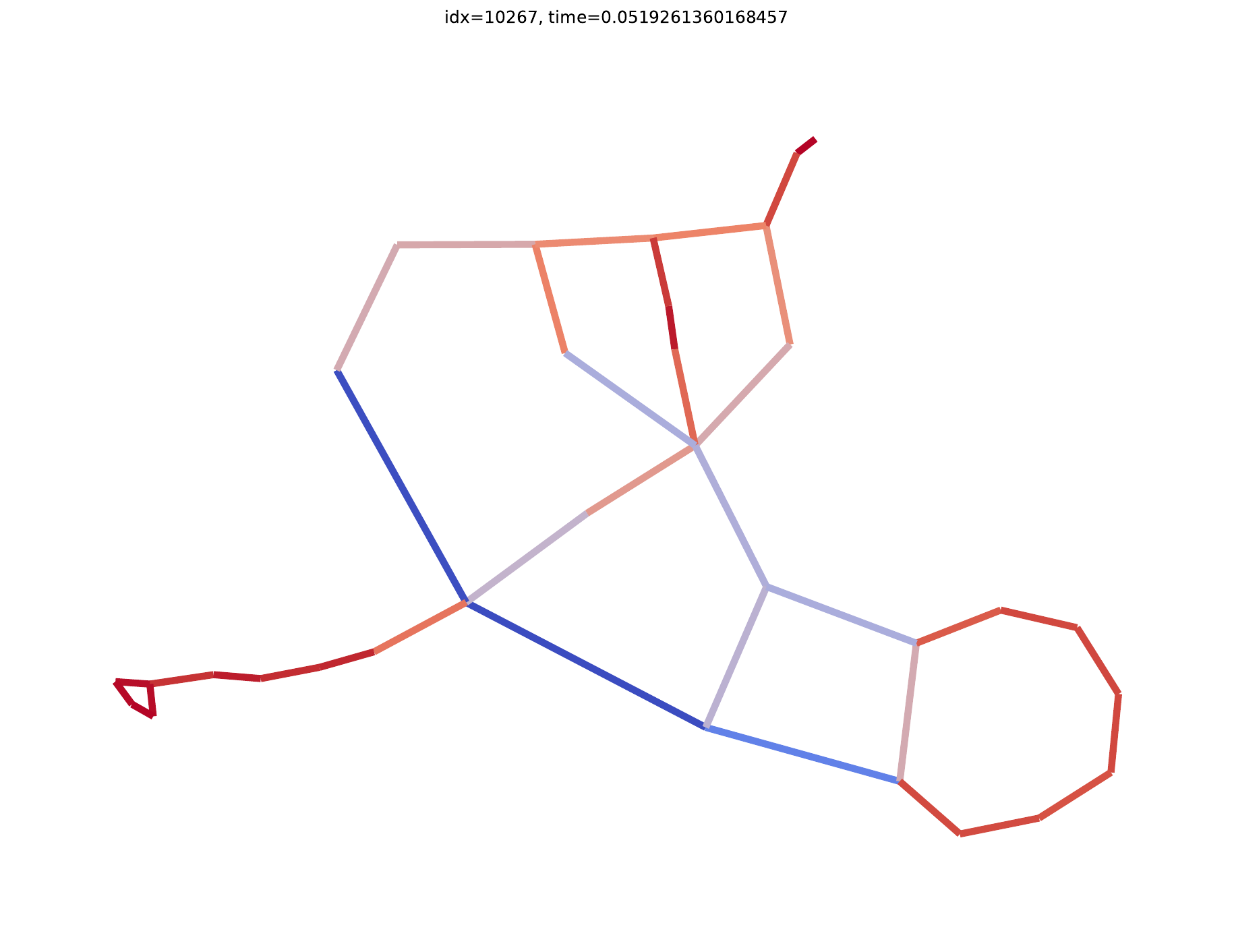} &
\imgcell{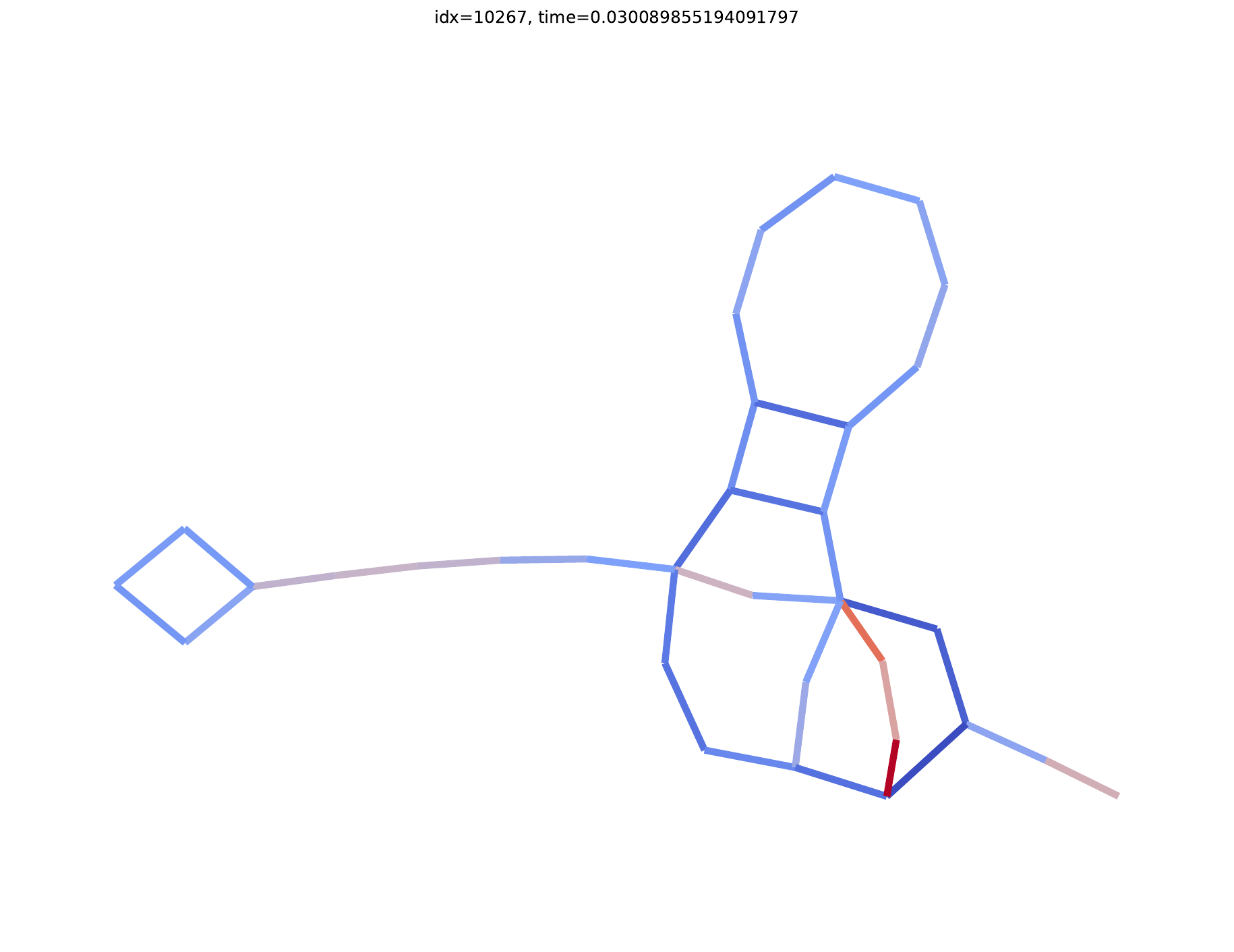} &
\imgcell{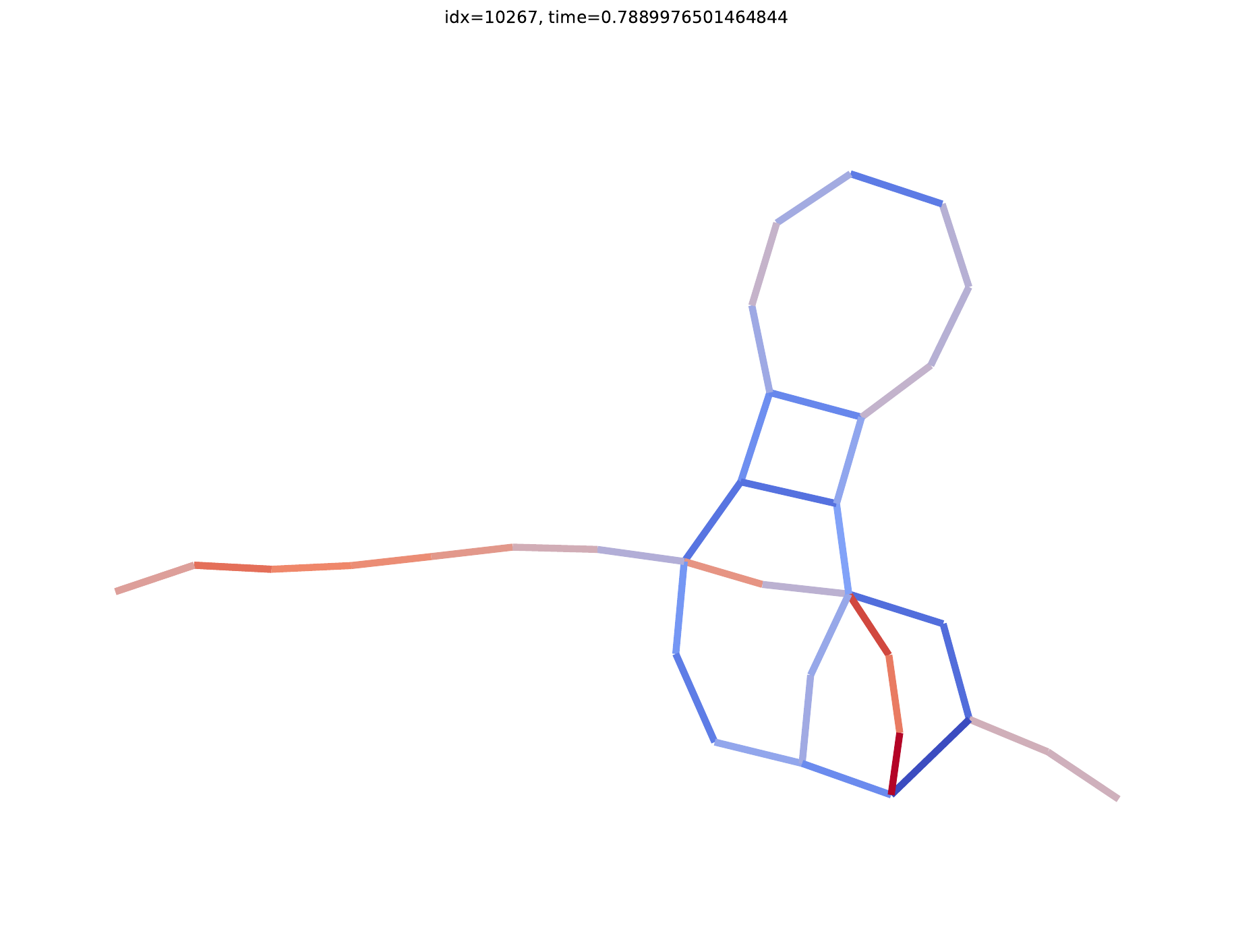} &
\imgcell{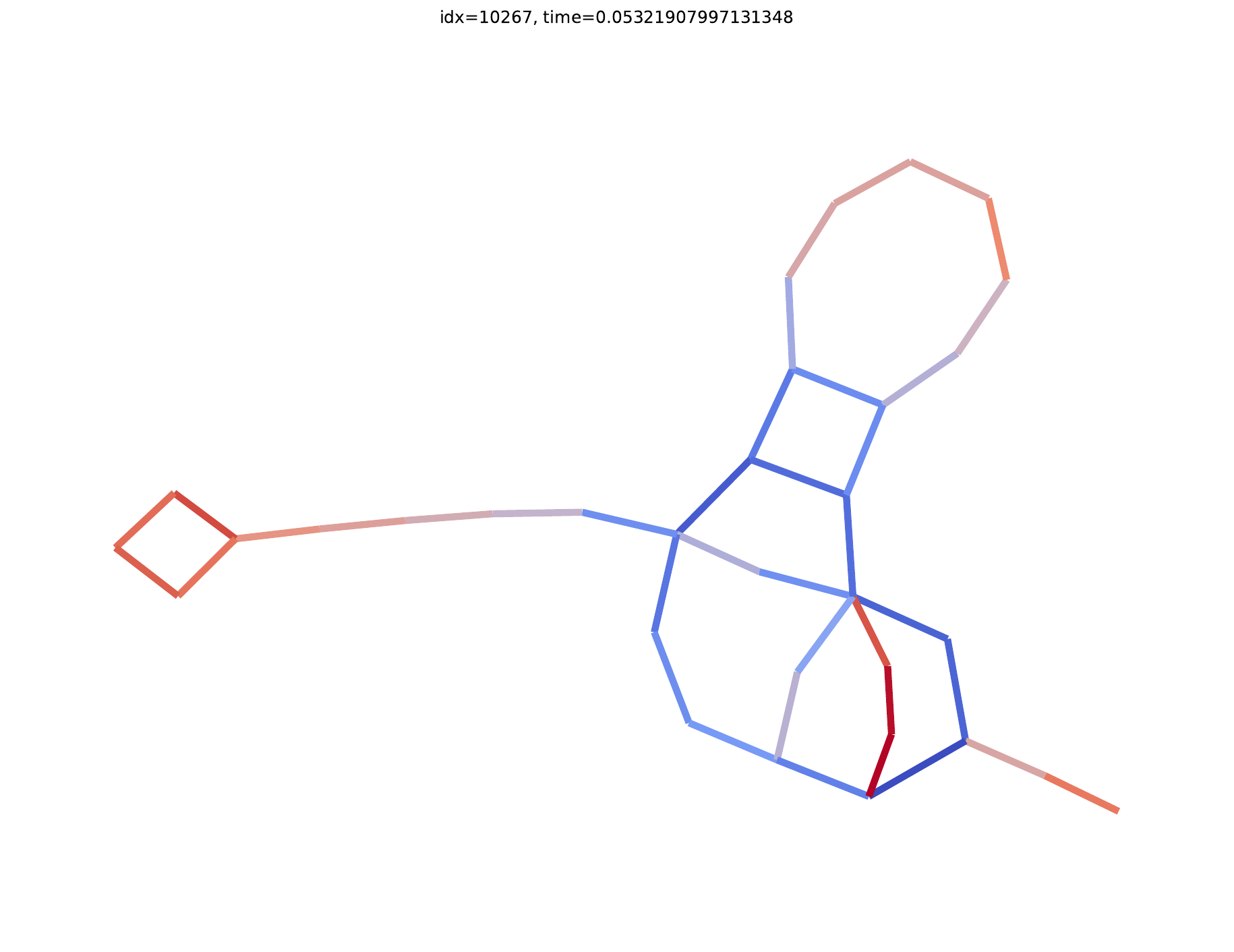} \\

&
t = 0.00s &
t = 0.31s &
t = 0.09s &
t = 0.05s &
t = 99.47s &
t = 0.03s &
t = 0.03s &
t = 0.04s &
t = 0.05s &
t = 0.03s &
t = 0.04s &
t = 0.05s \\

\makecell{\bfseries grafo2968.32\\N = 100\\M = 136} &
\imgcell{figures/rome_graphs/10277_sgd2.pdf} &
\imgcell{figures/rome_graphs/10277_pmds.pdf} &
\imgcell{figures/rome_graphs/10277_fa2.pdf} &
\imgcell{figures/rome_graphs/10277_deepgd.pdf} &
\imgcell{figures/rome_graphs/10277_gd2_stress_xing.pdf} &
\imgcell{figures/rome_graphs/10277_smartgd_stress.pdf} &
\imgcell{figures/rome_graphs/10277_smartgd_xing.pdf} &
\imgcell{figures/rome_graphs/10277_smartgd_xing_nsc.pdf} &
\imgcell{figures/rome_graphs/10277_smartgd_xangle.pdf} &
\imgcell{figures/rome_graphs/10277_smartgd_stress_xing.pdf} &
\imgcell{figures/rome_graphs/10277_smartgd_stress_xangle.pdf} &
\imgcell{figures/rome_graphs/10277_smartgd_combined.pdf} \\

&
t = 0.00s &
t = 0.34s &
t = 0.09s &
t = 0.04s &
t = 101.28s &
t = 0.02s &
t = 0.02s &
t = 0.02s &
t = 0.02s &
t = 0.02s &
t = 0.02s &
t = 0.02s \\

\makecell{\bfseries grafo2860.18\\N = 35\\M = 38} &
\imgcell{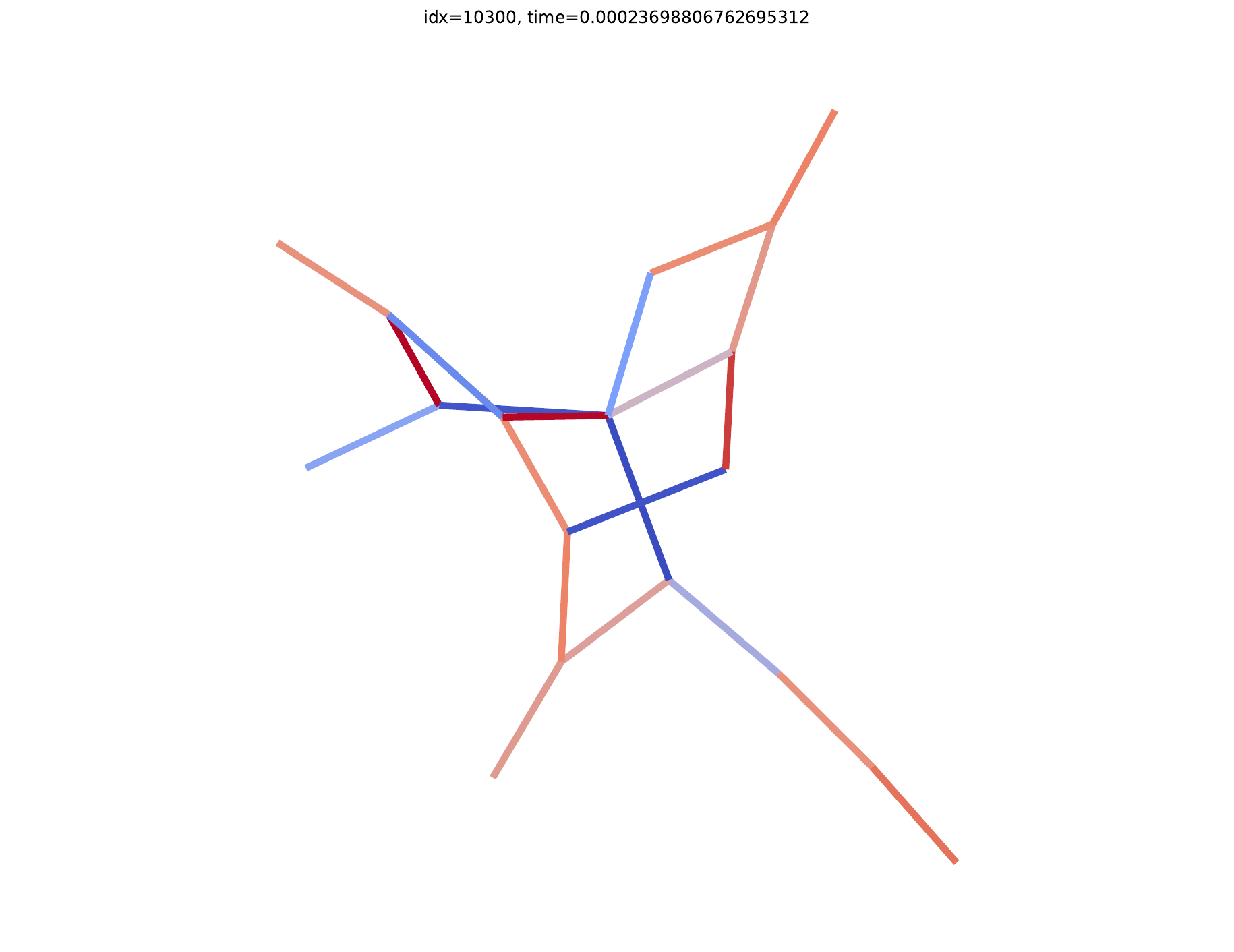} &
\imgcell{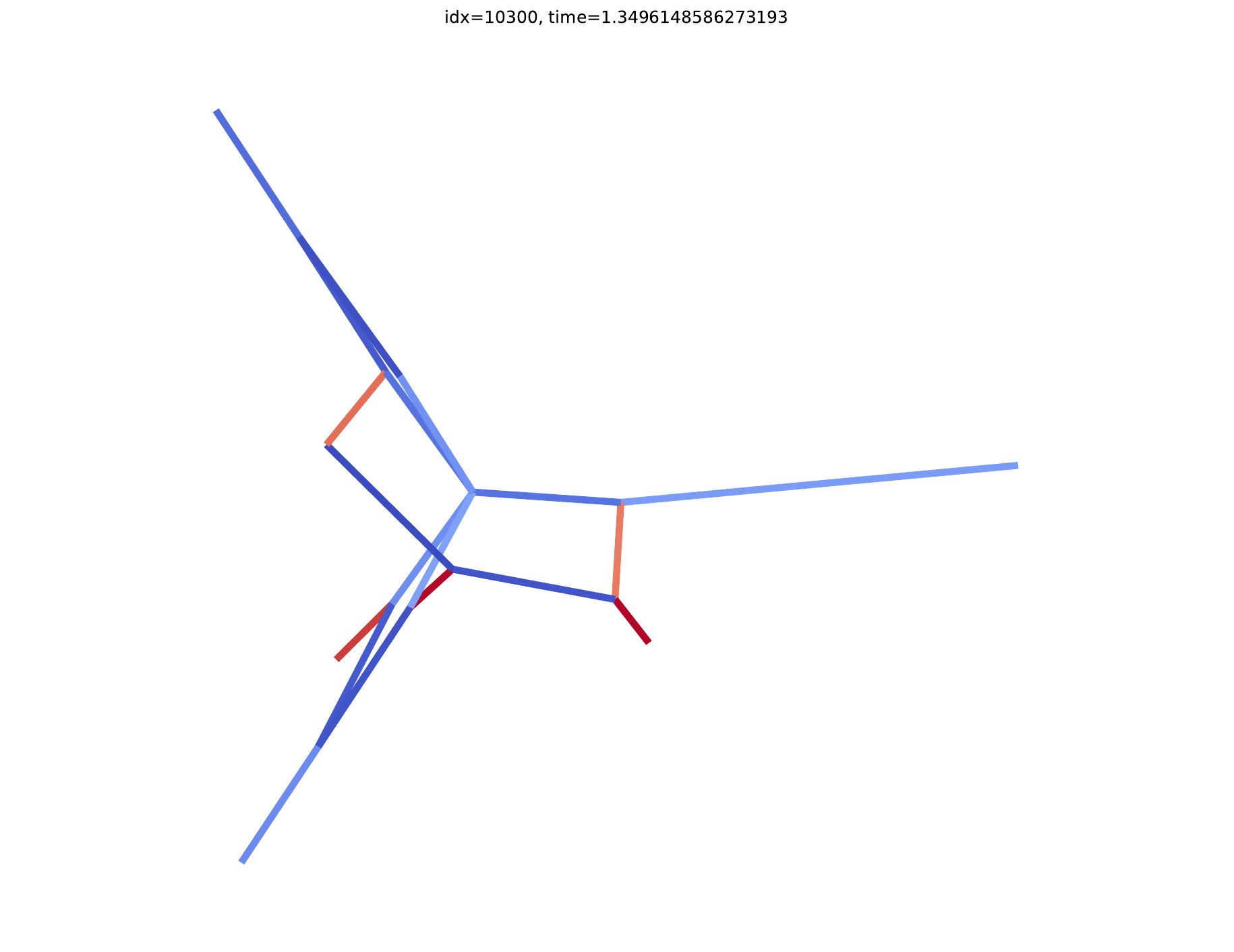} &
\imgcell{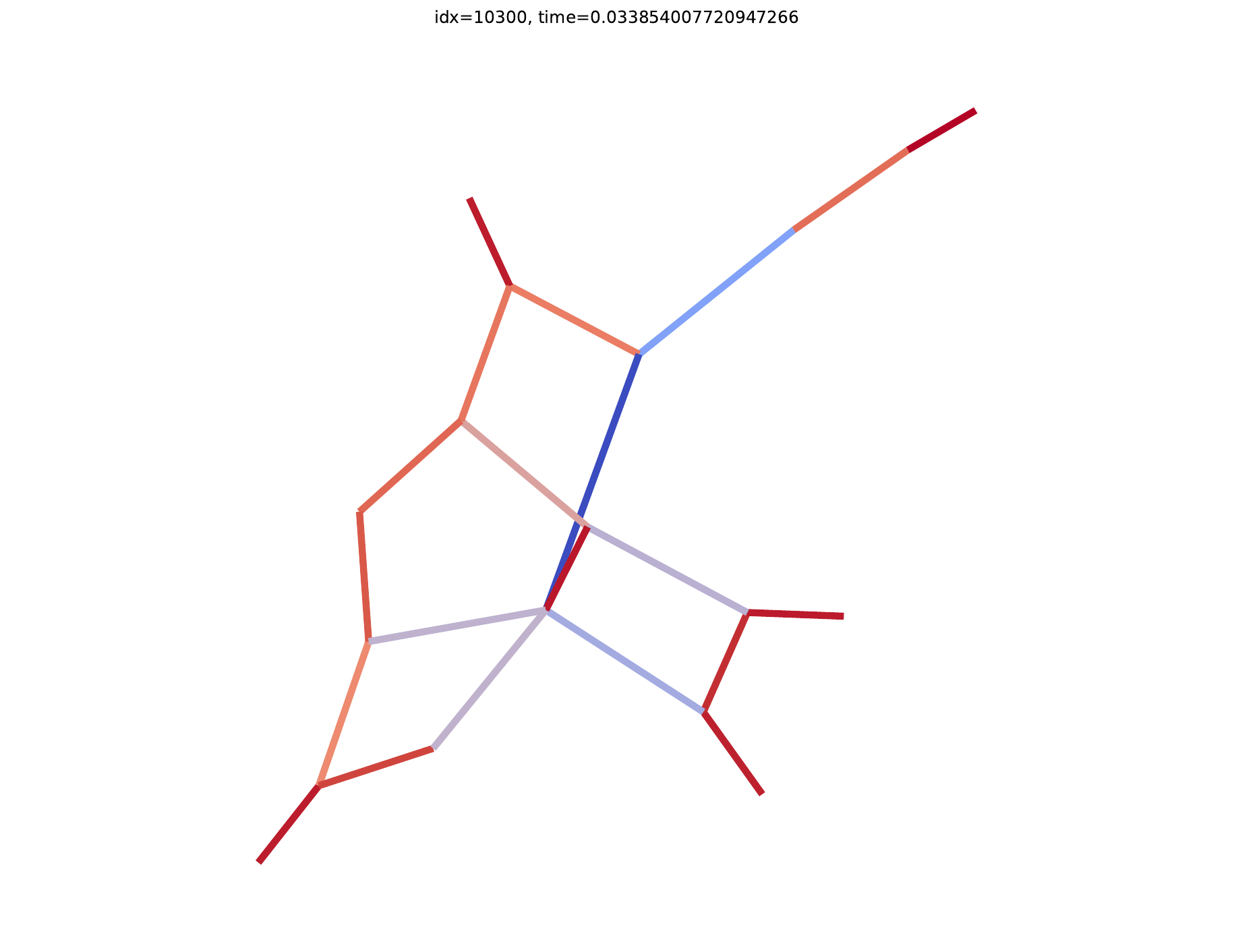} &
\imgcell{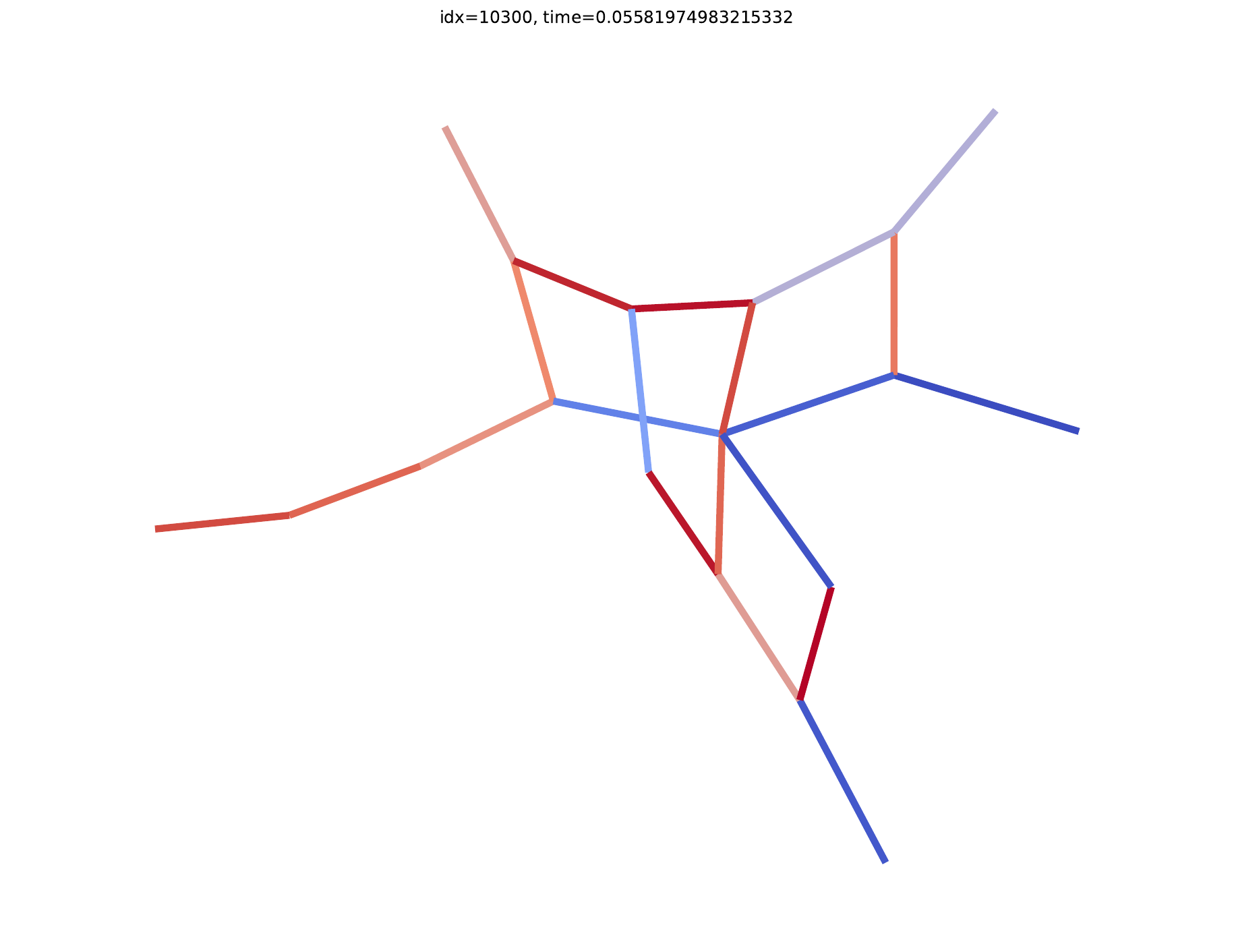} &
\imgcell{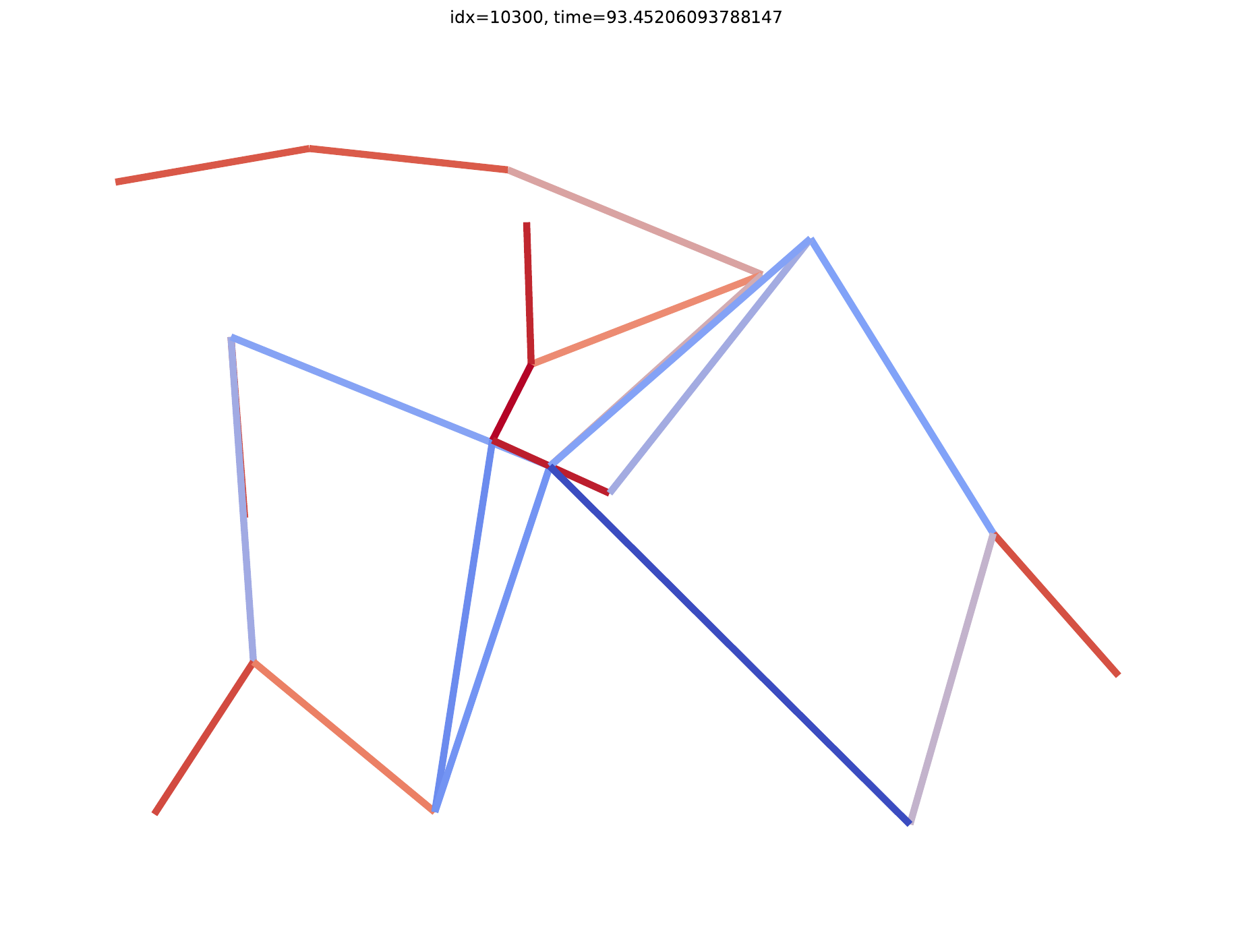} &
\imgcell{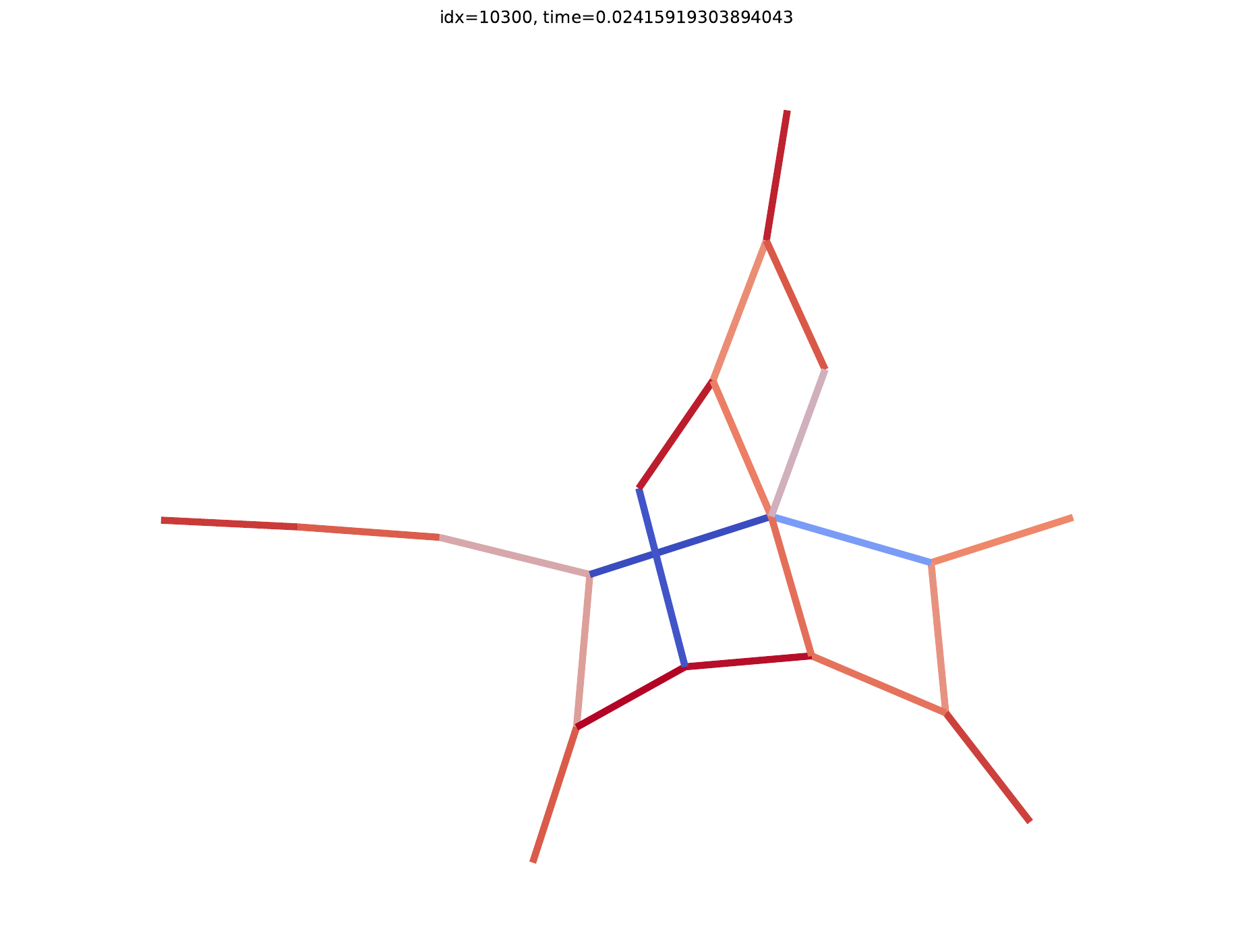} &
\imgcell{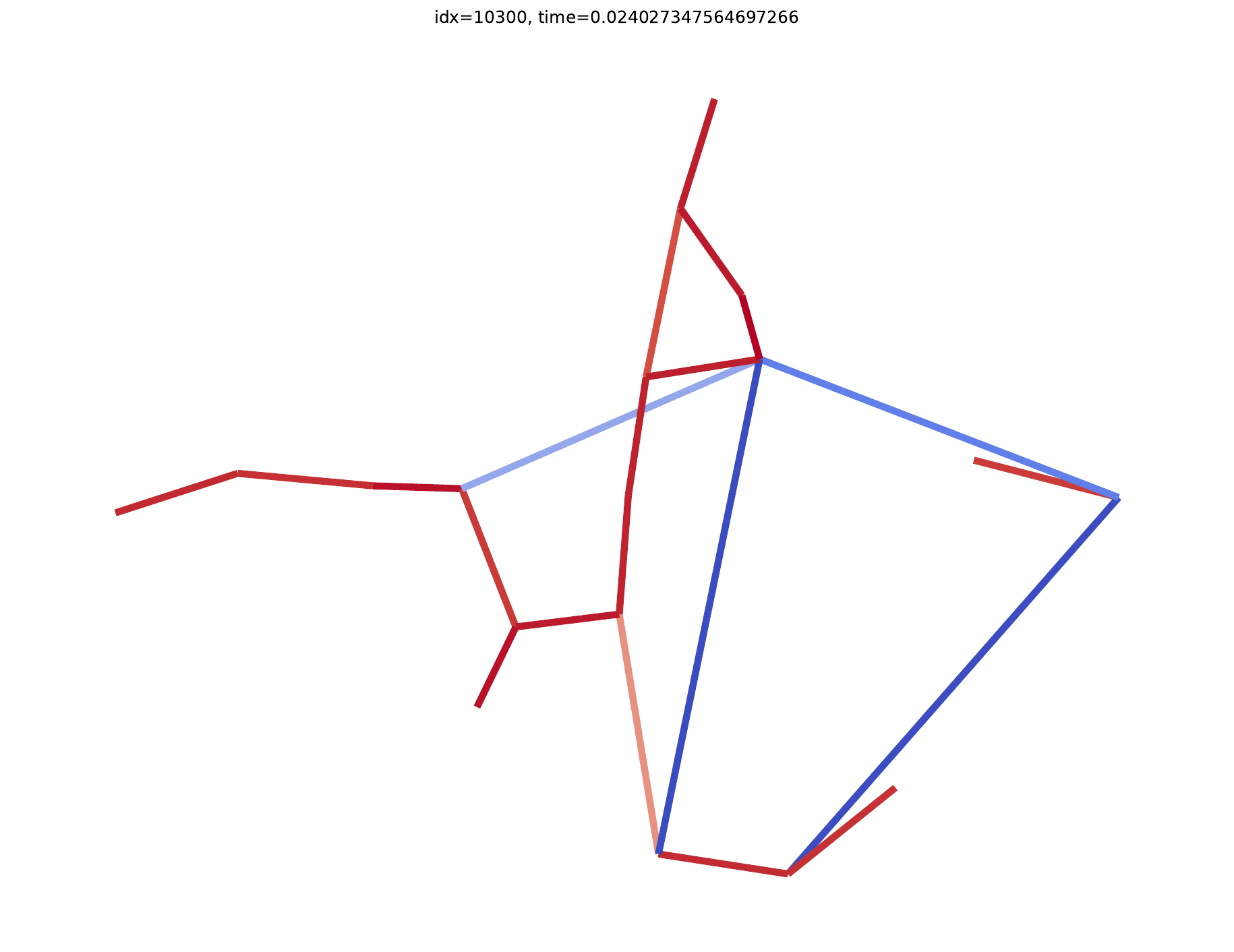} &
\imgcell{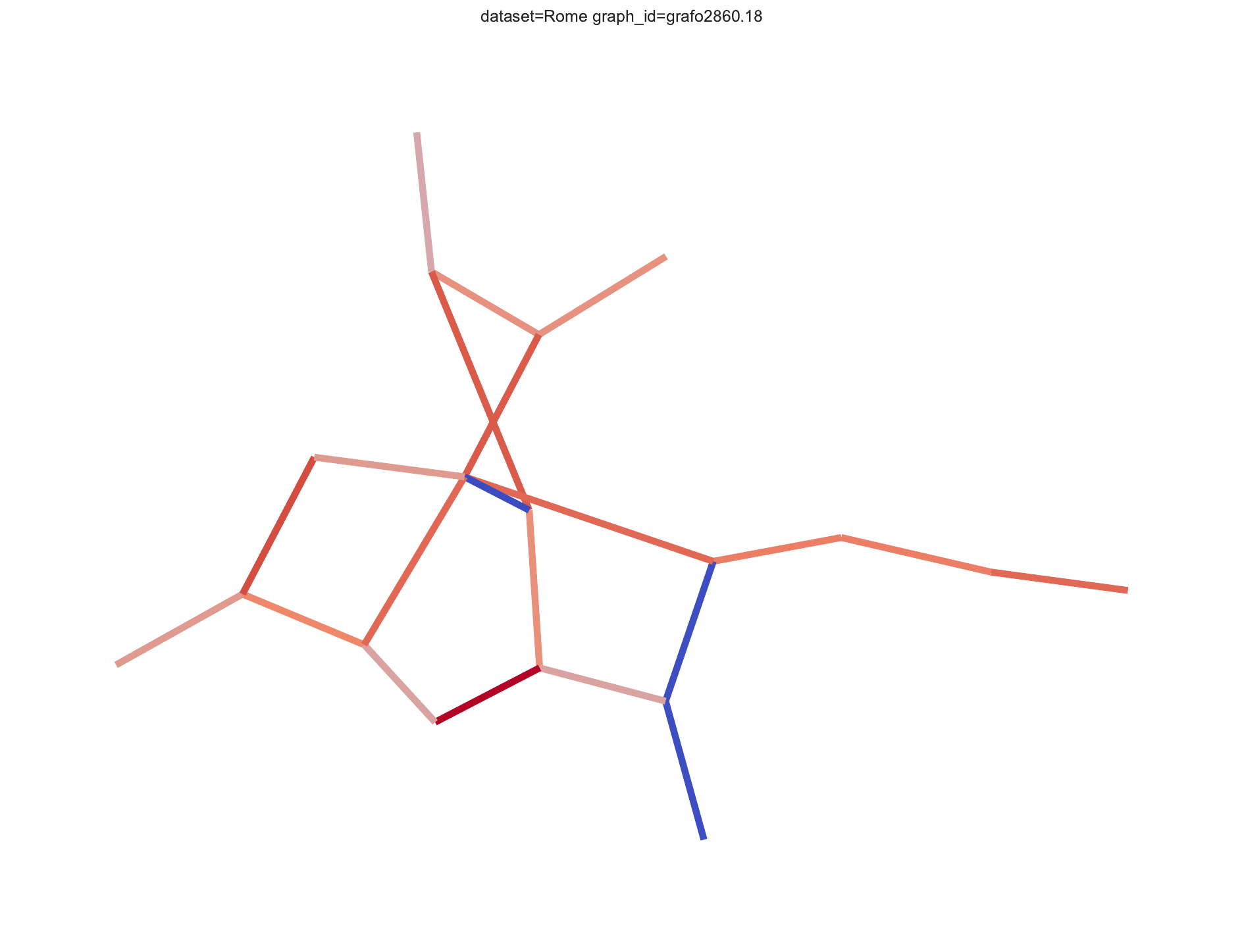} &
\imgcell{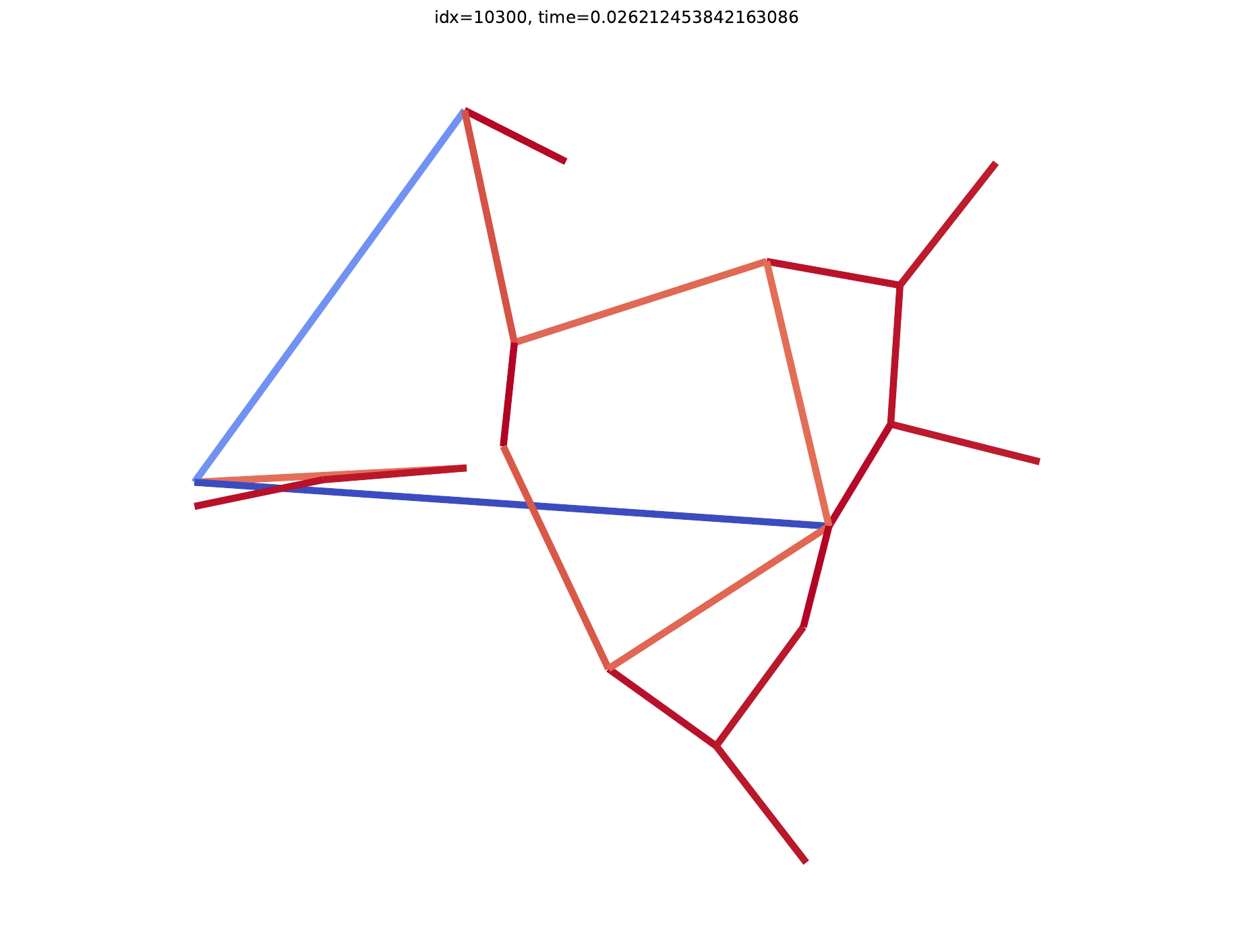} &
\imgcell{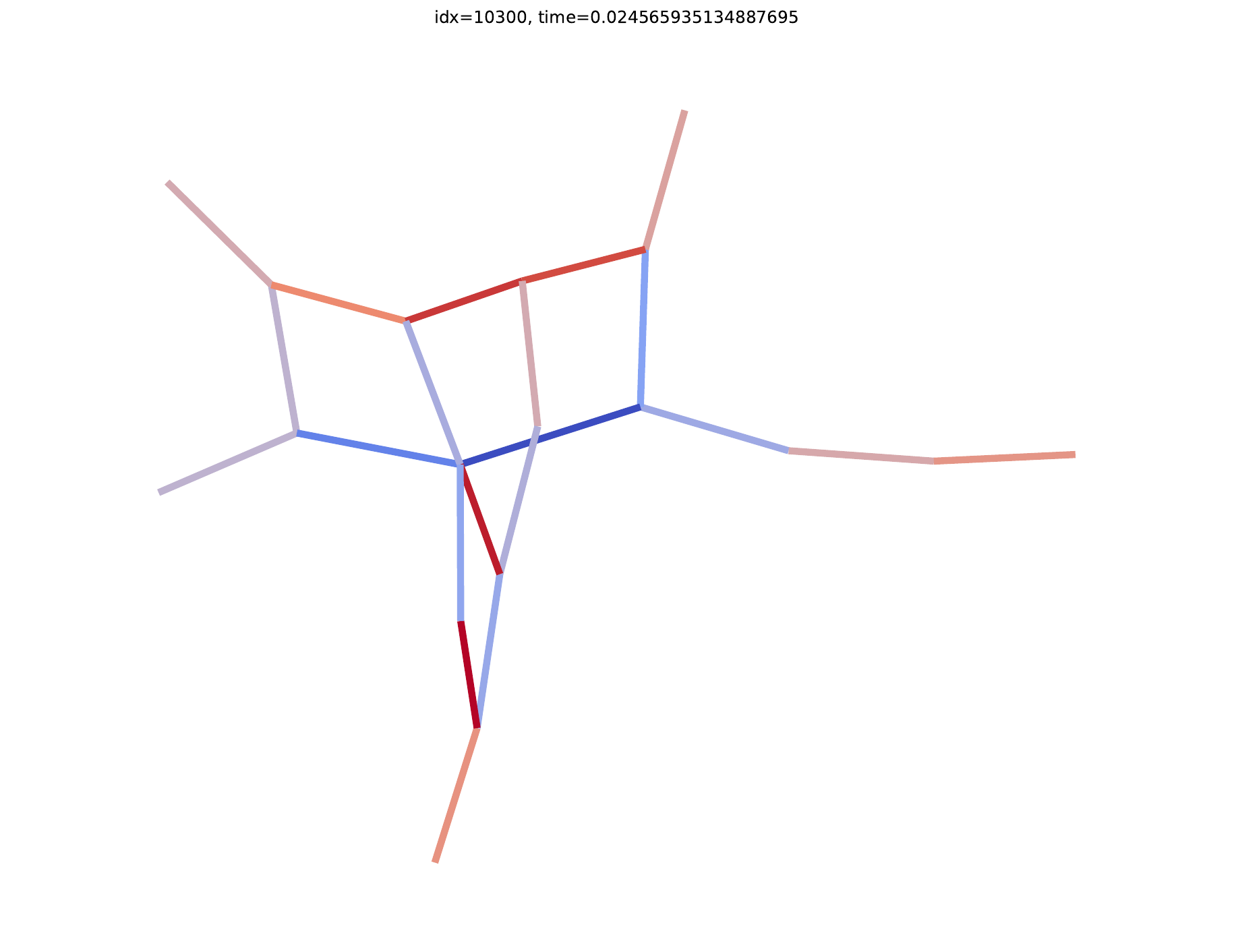} &
\imgcell{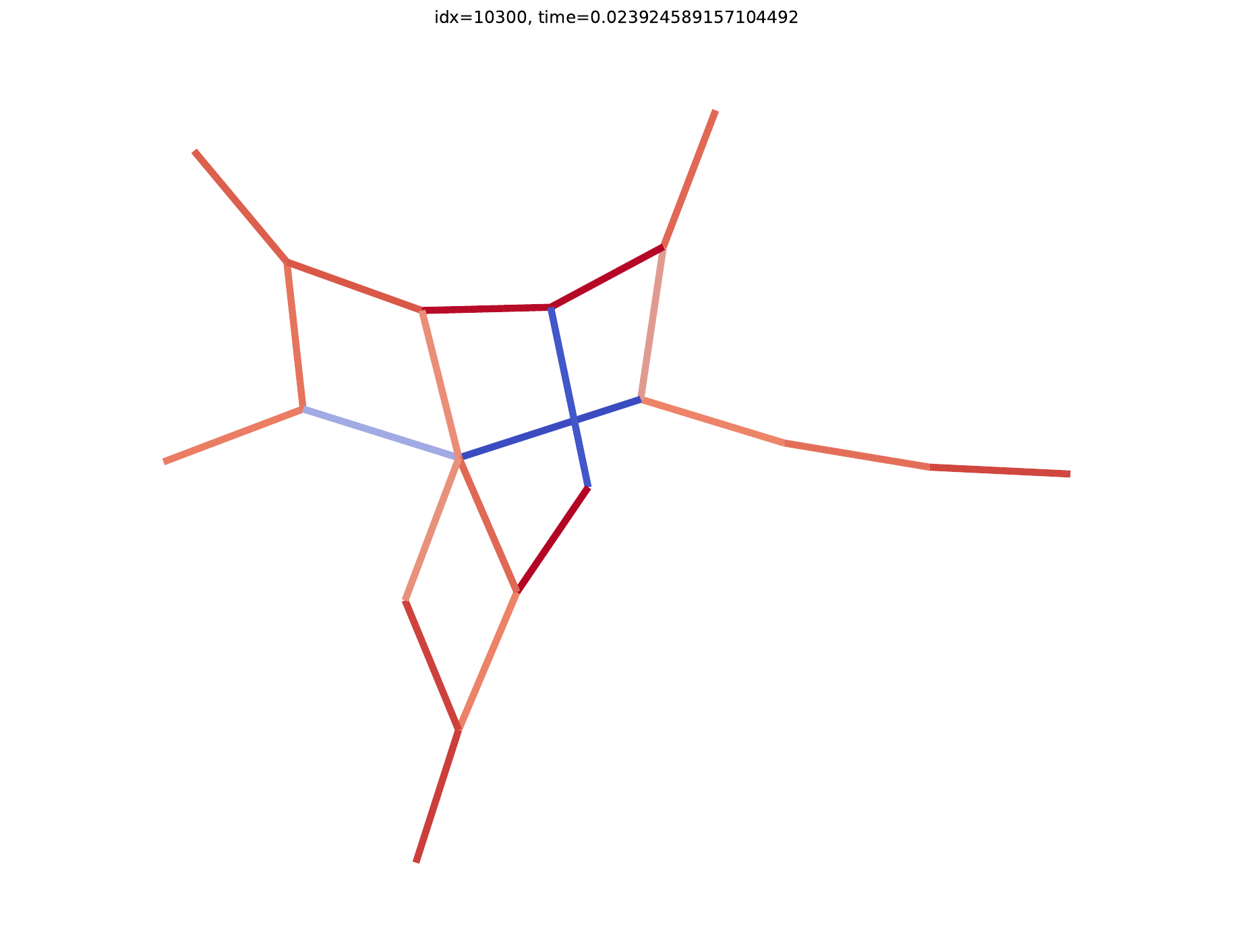} &
\imgcell{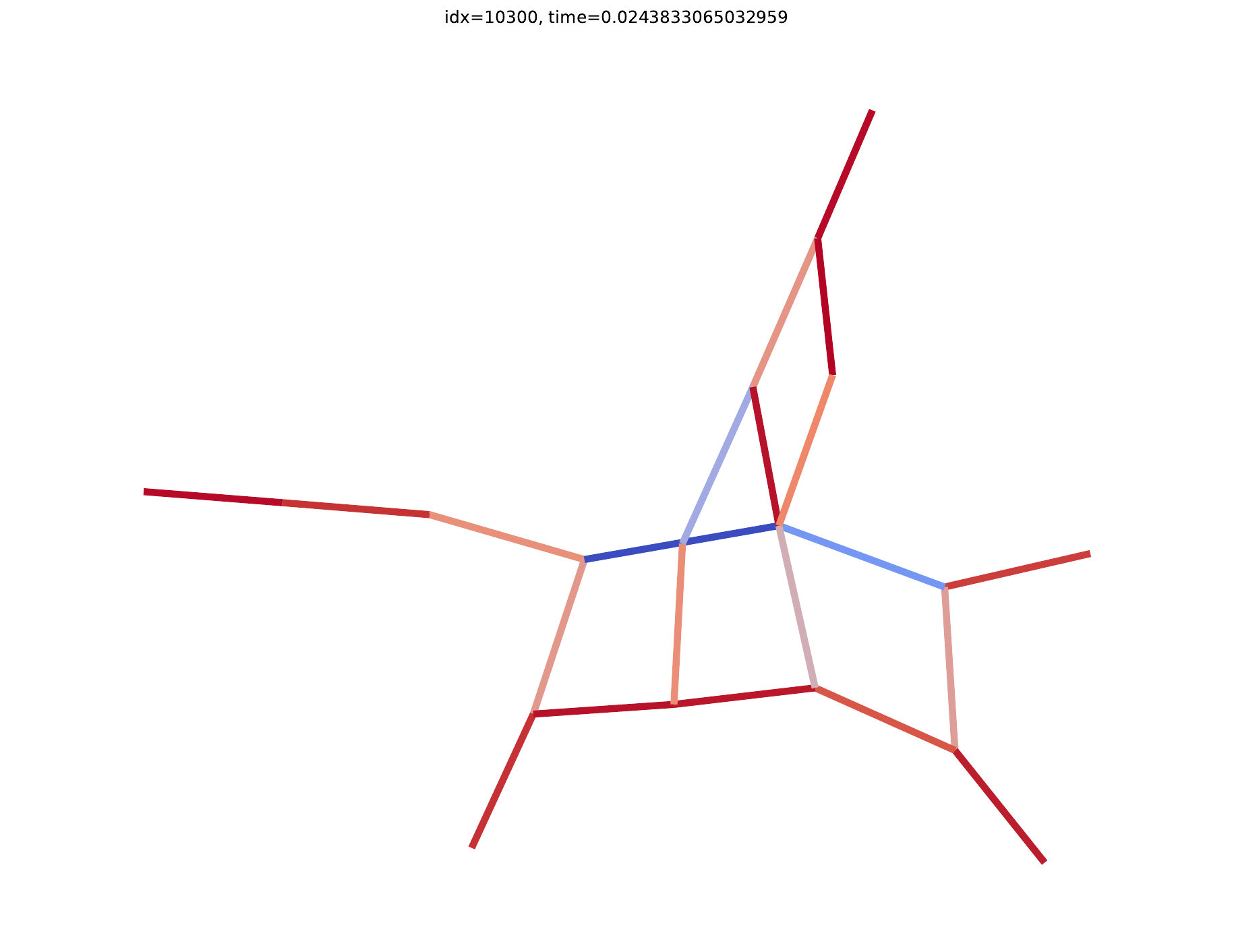} \\

&
t = 0.00s &
t = 1.35s &
t = 0.03s &
t = 0.06s &
t = 93.45s &
t = 0.02s &
t = 0.02s &
t = 0.02s &
t = 0.03s &
t = 0.02s &
t = 0.02s &
t = 0.02s \\

\makecell{\bfseries grafo5262.37\\N = 83\\M = 117} &
\imgcell{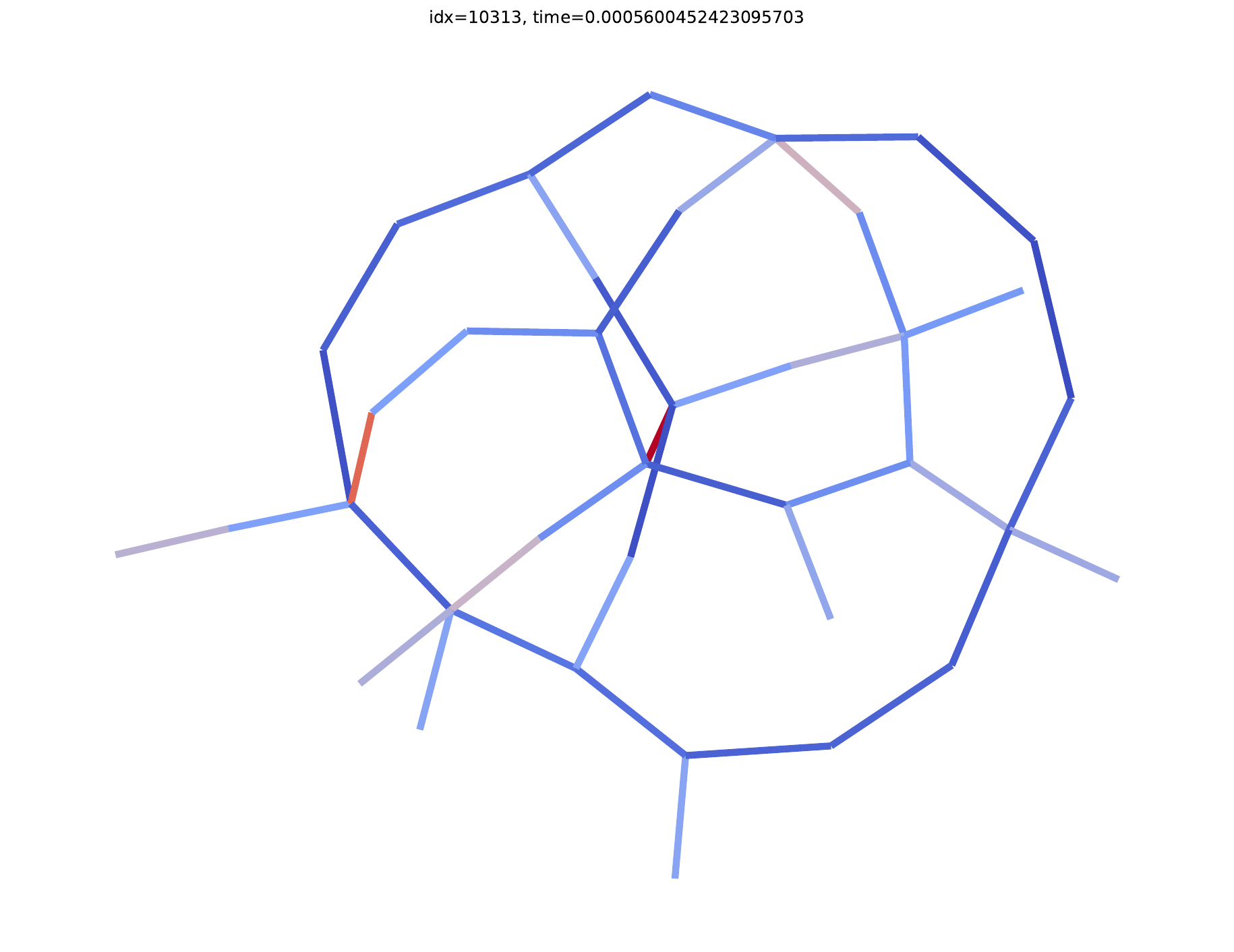} &
\imgcell{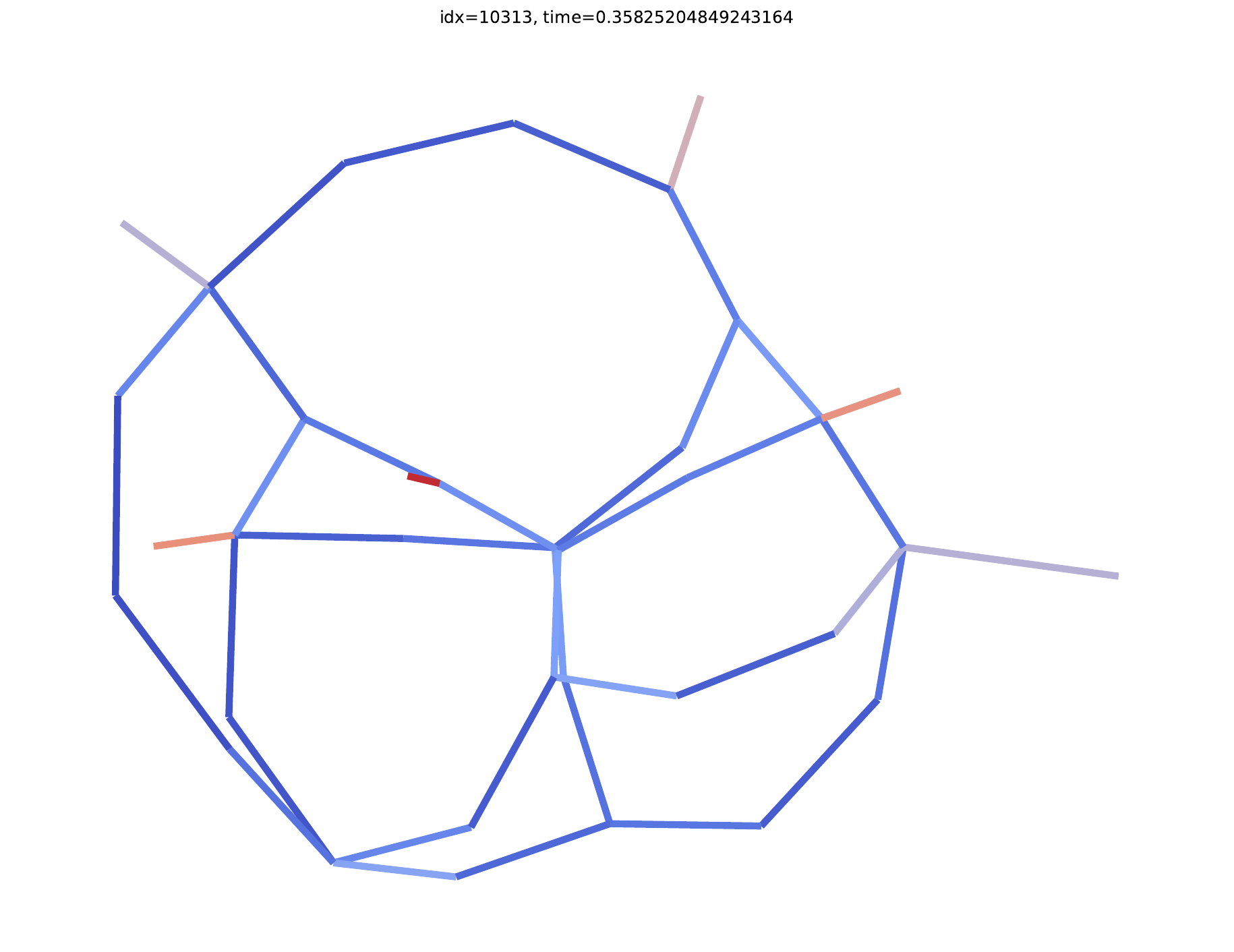} &
\imgcell{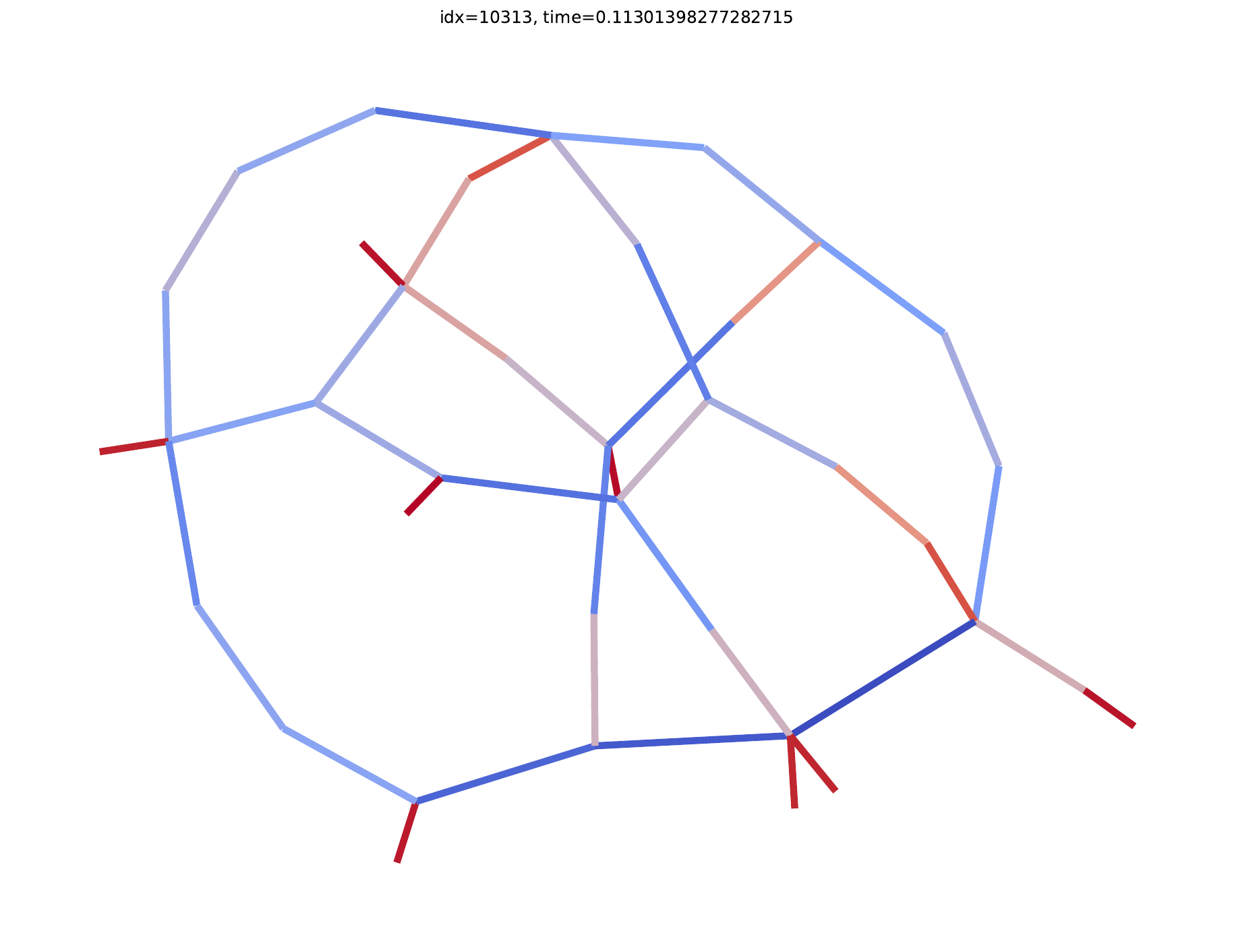} &
\imgcell{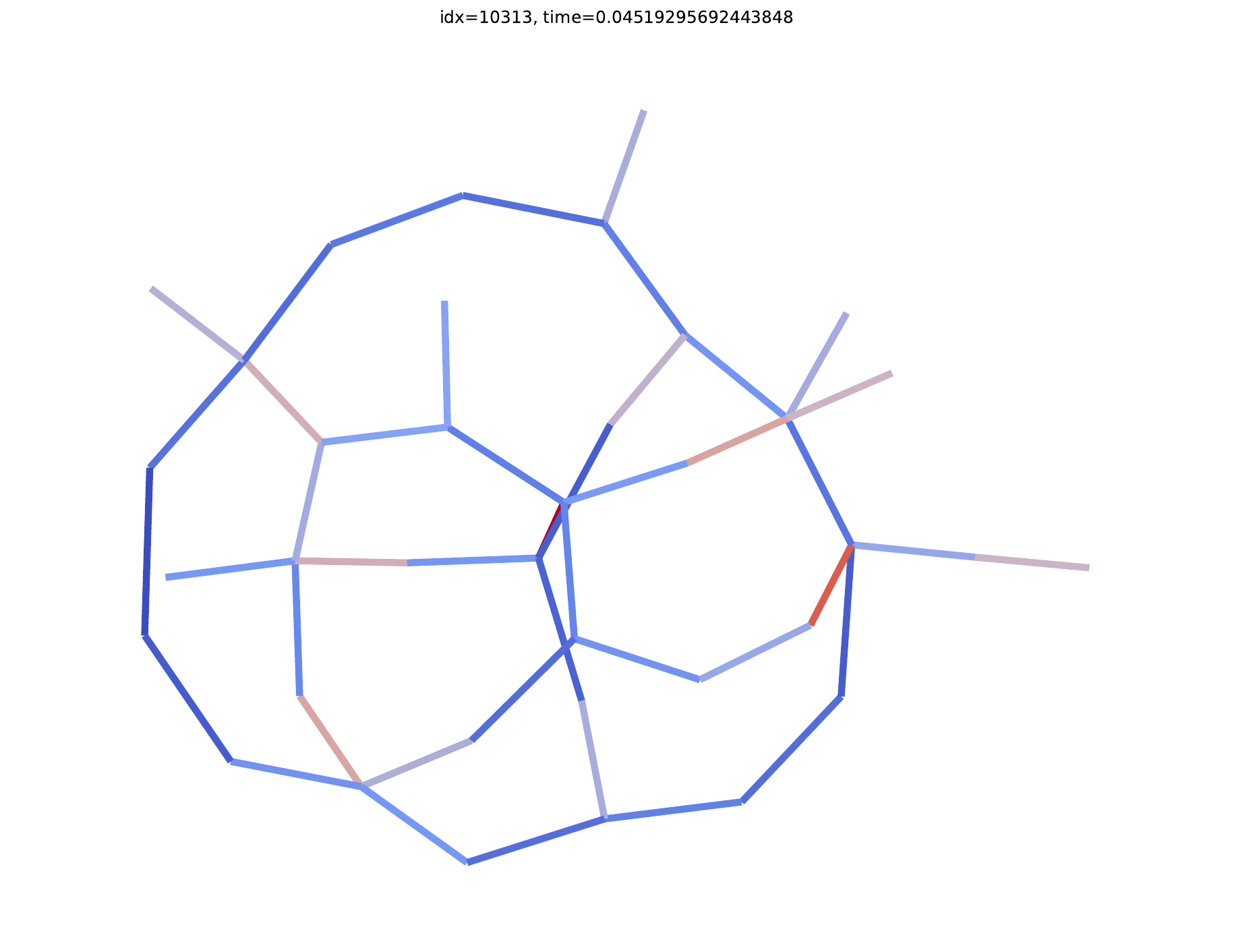} &
\imgcell{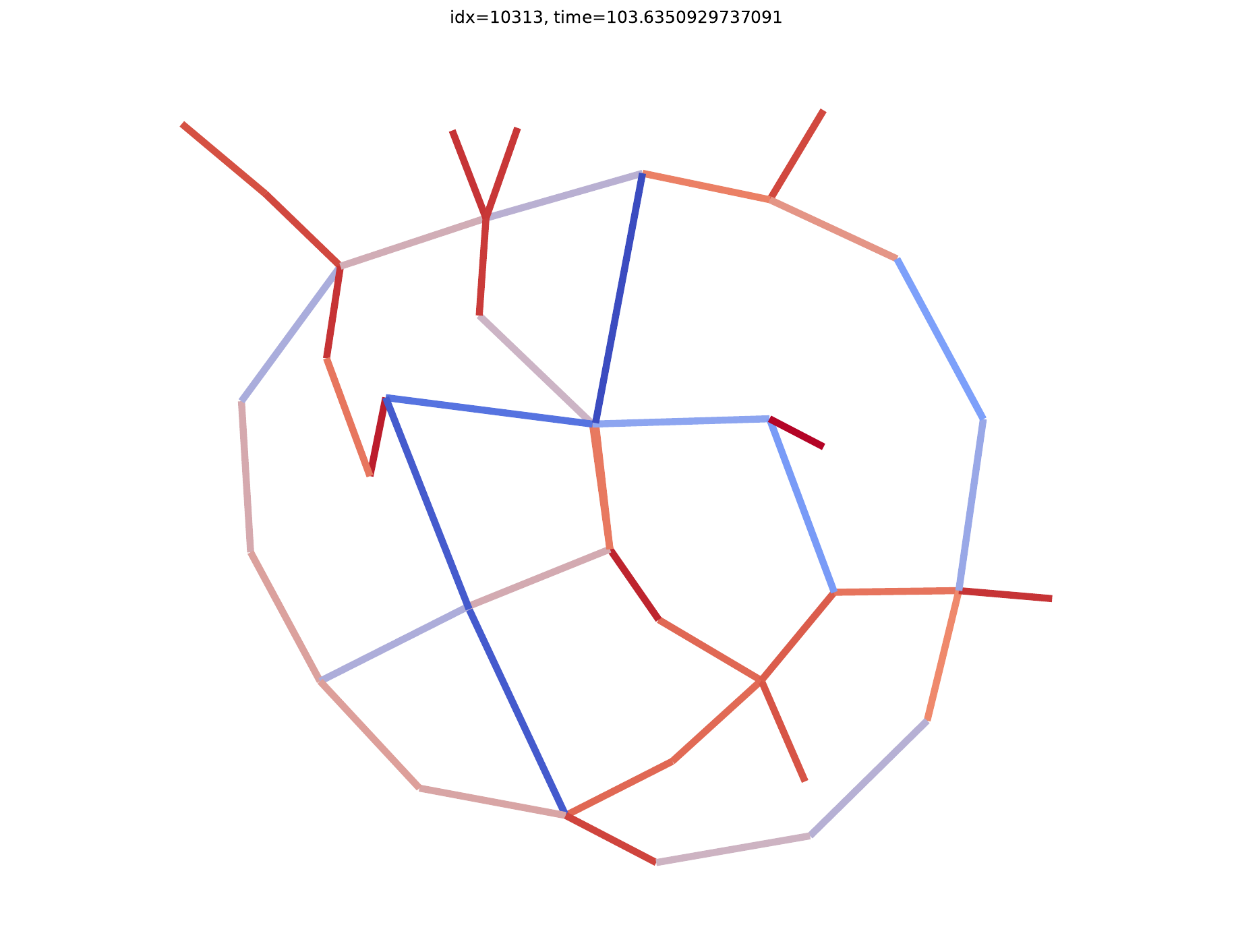} &
\imgcell{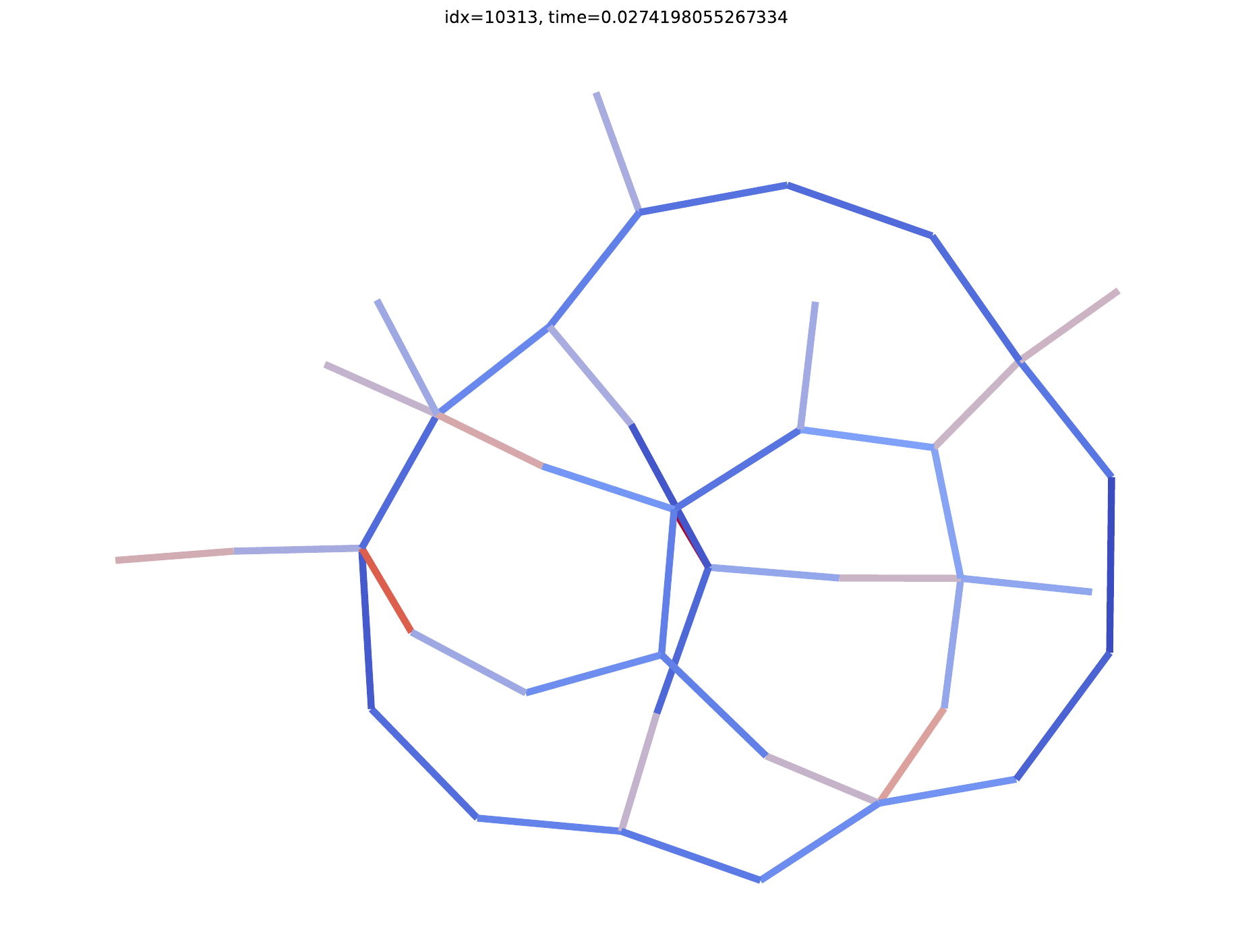} &
\imgcell{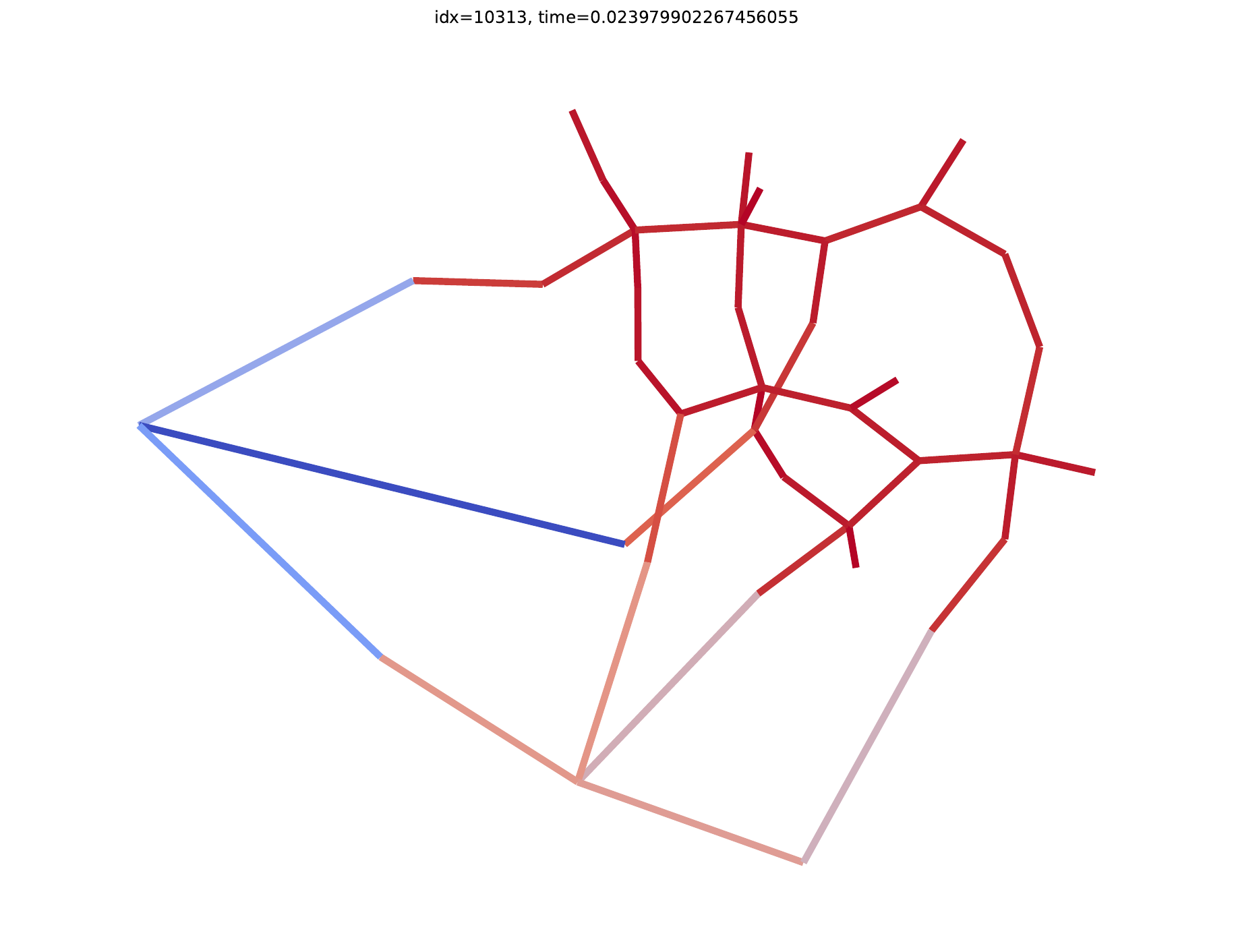} &
\imgcell{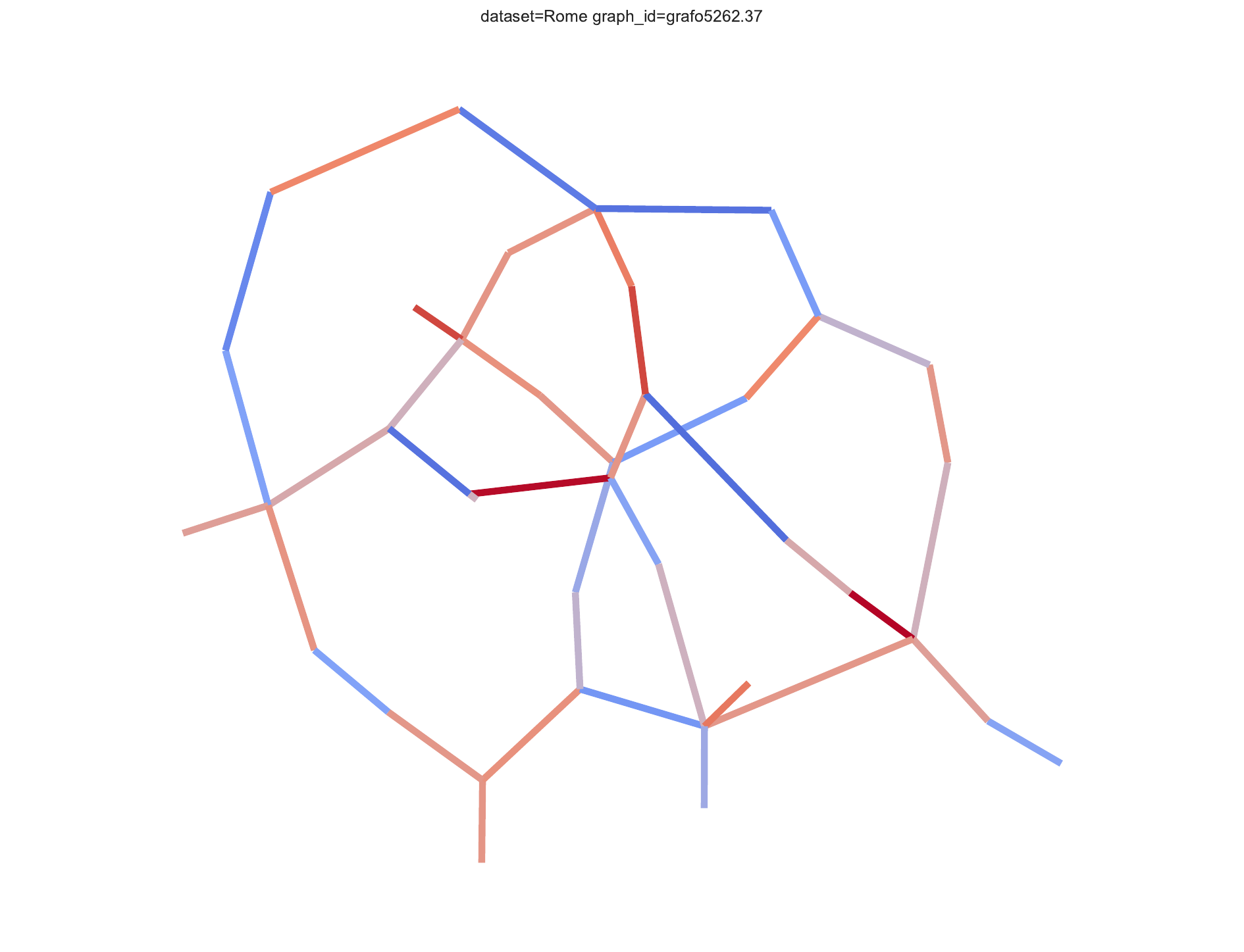} &
\imgcell{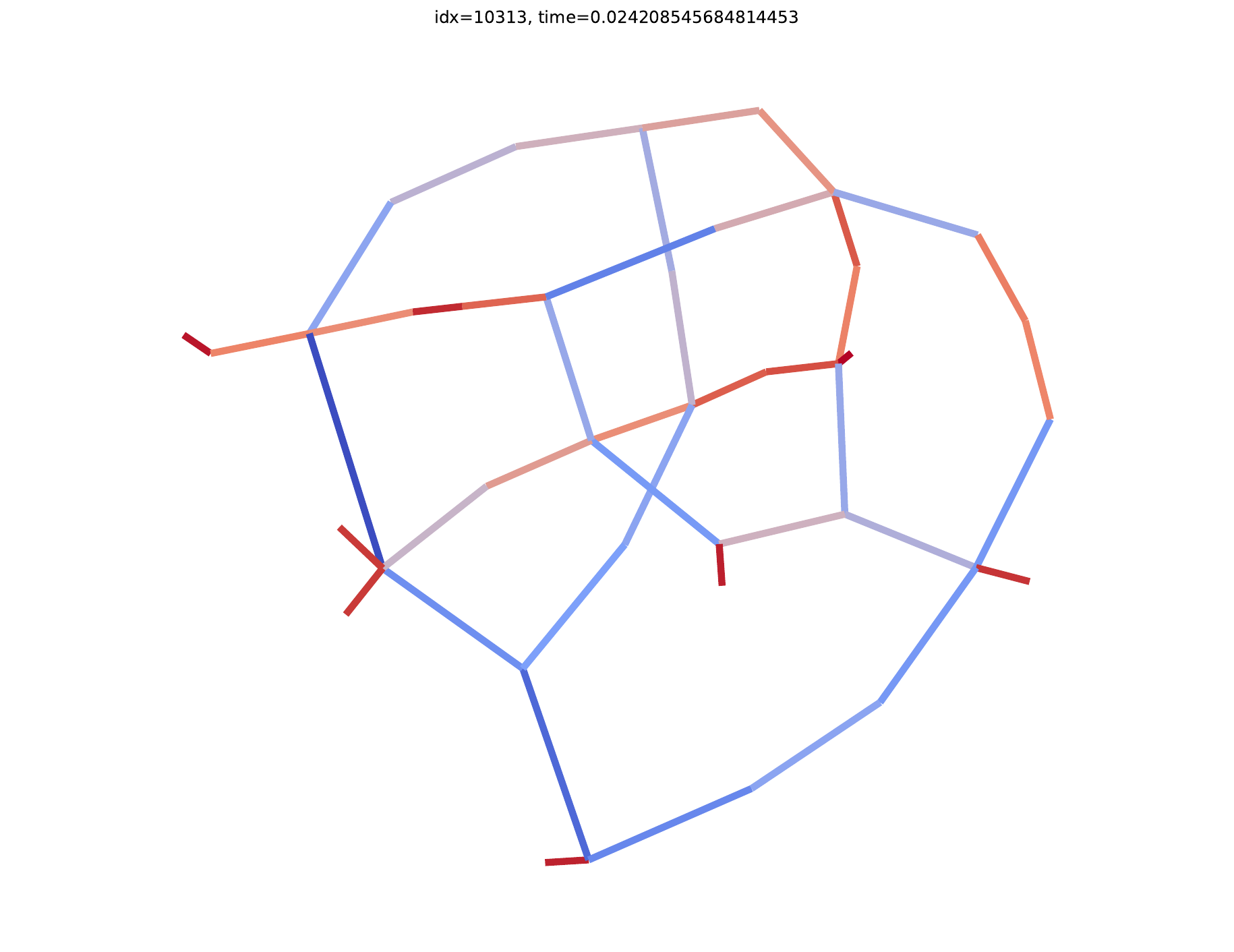} &
\imgcell{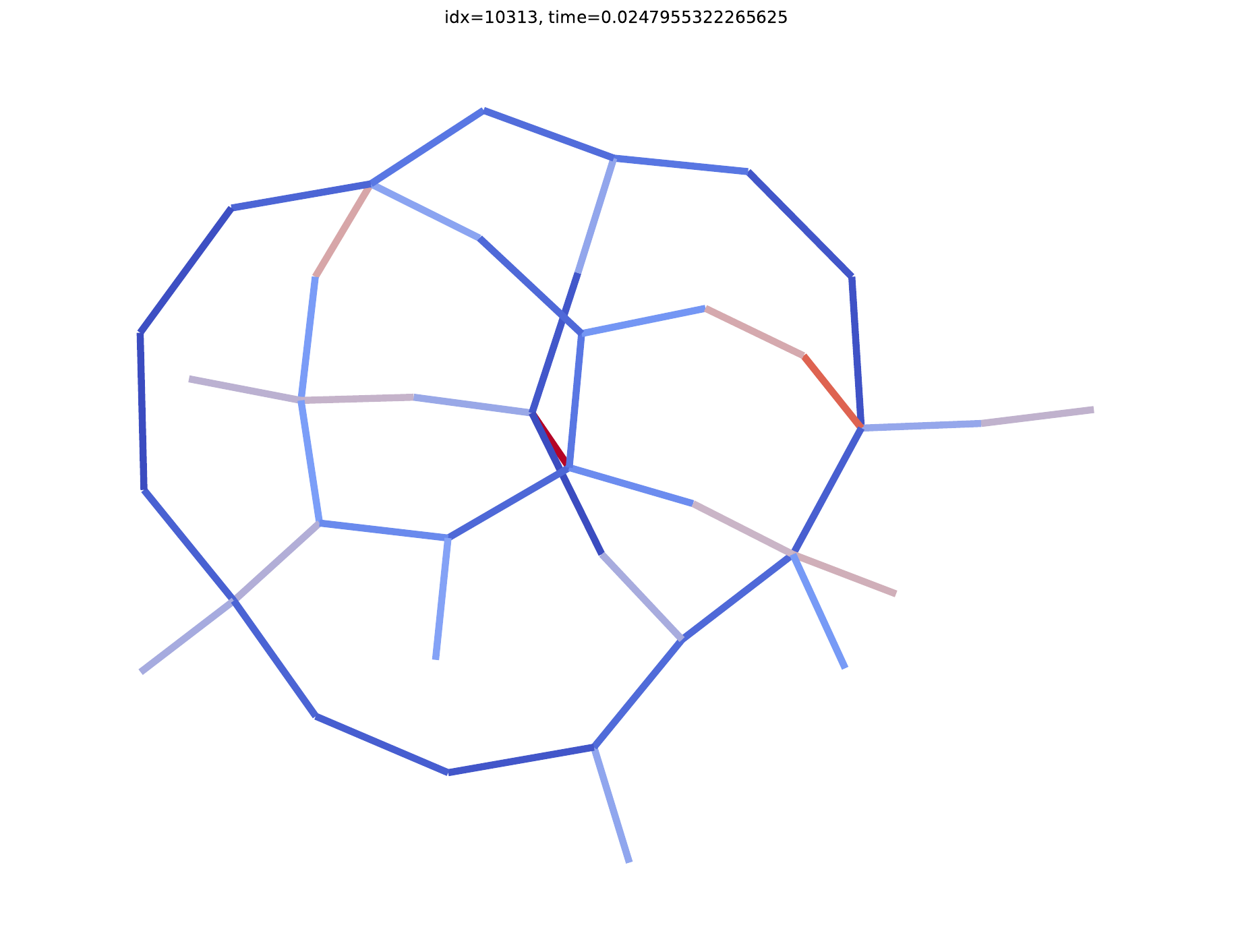} &
\imgcell{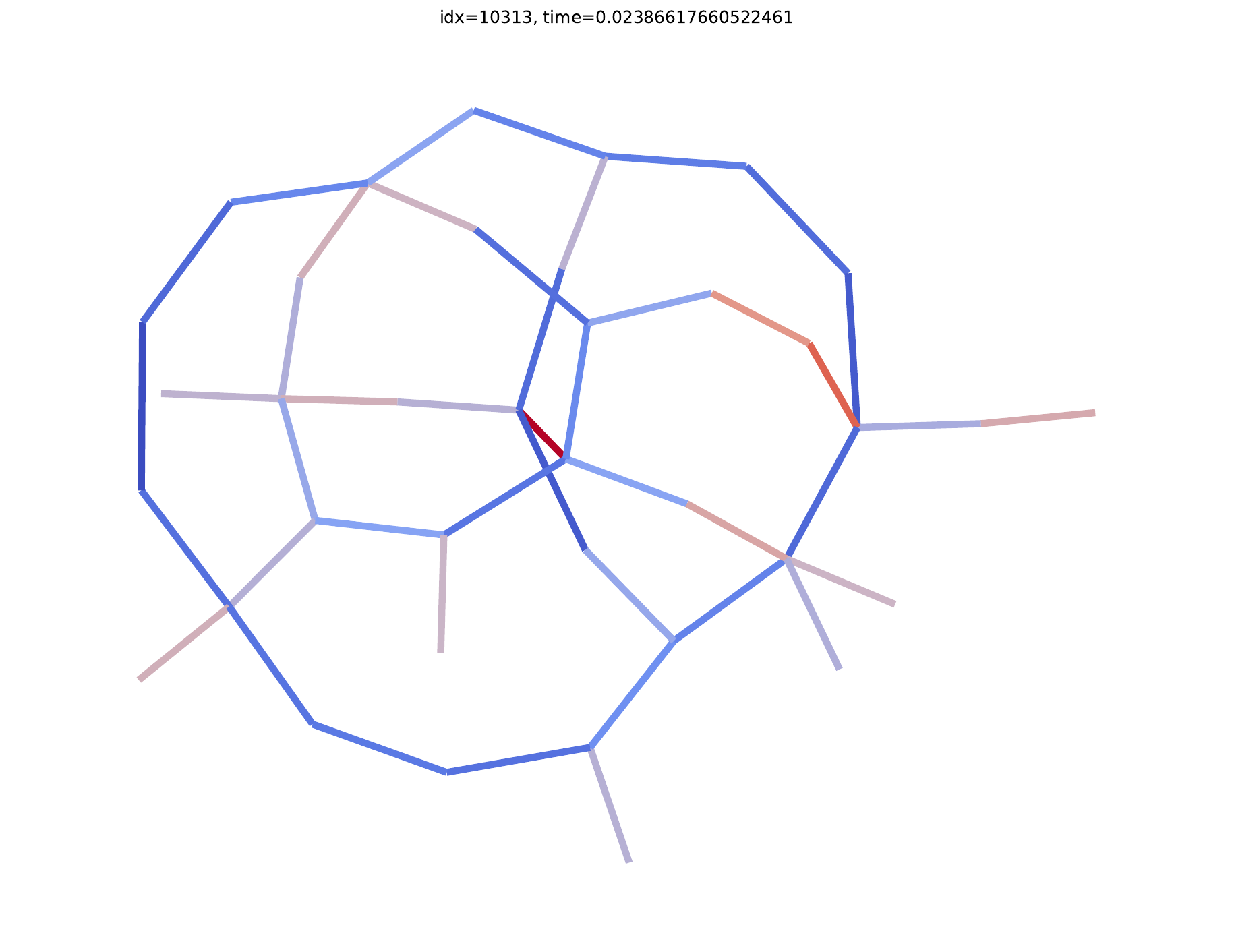} &
\imgcell{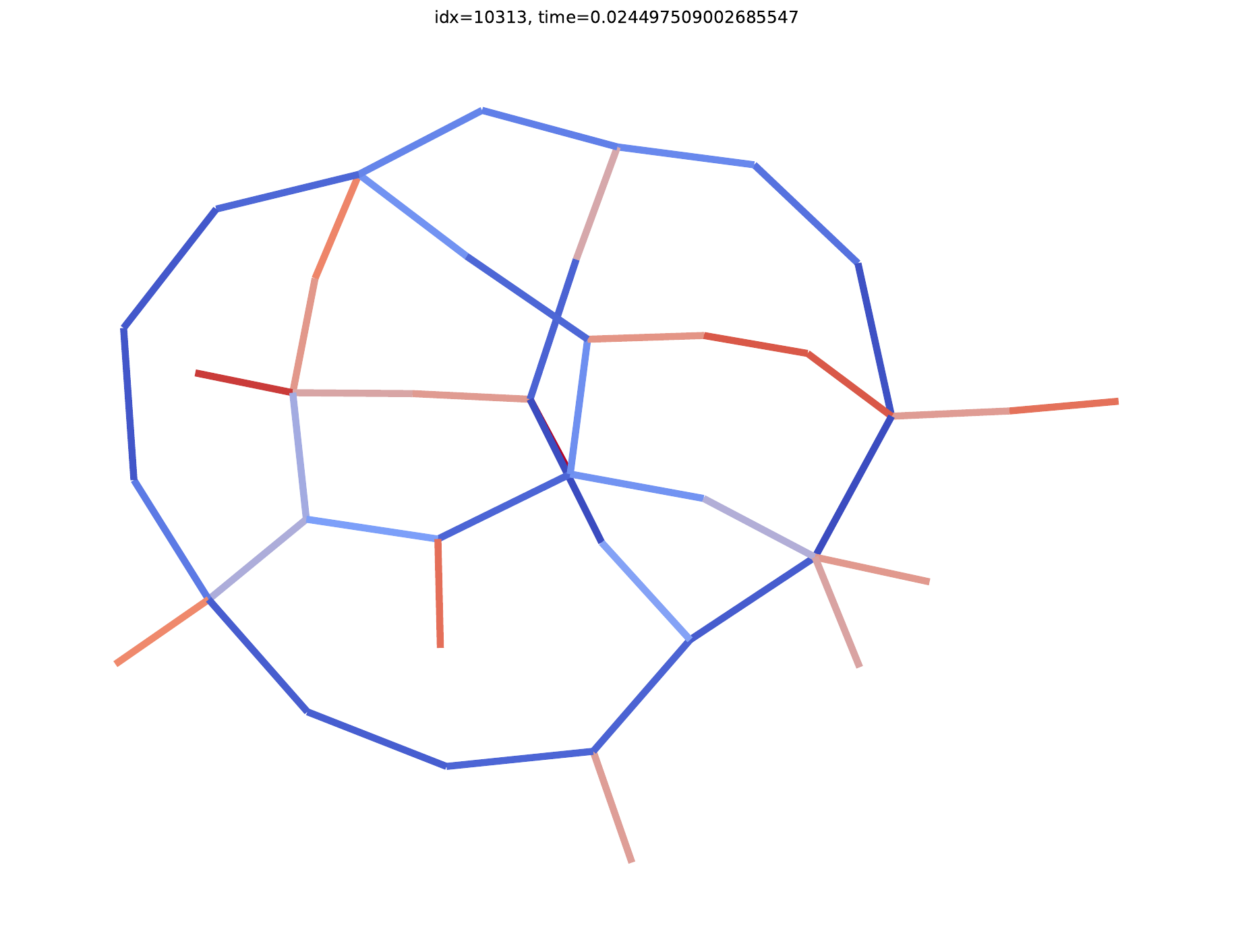} \\

&
t = 0.00s &
t = 0.36s &
t = 0.11s &
t = 0.05s &
t = 103.64s &
t = 0.03s &
t = 0.02s &
t = 0.02s &
t = 0.02s &
t = 0.02s &
t = 0.02s &
t = 0.02s \\

\makecell{\bfseries grafo520.14\\N = 23\\M = 24} &
\imgcell{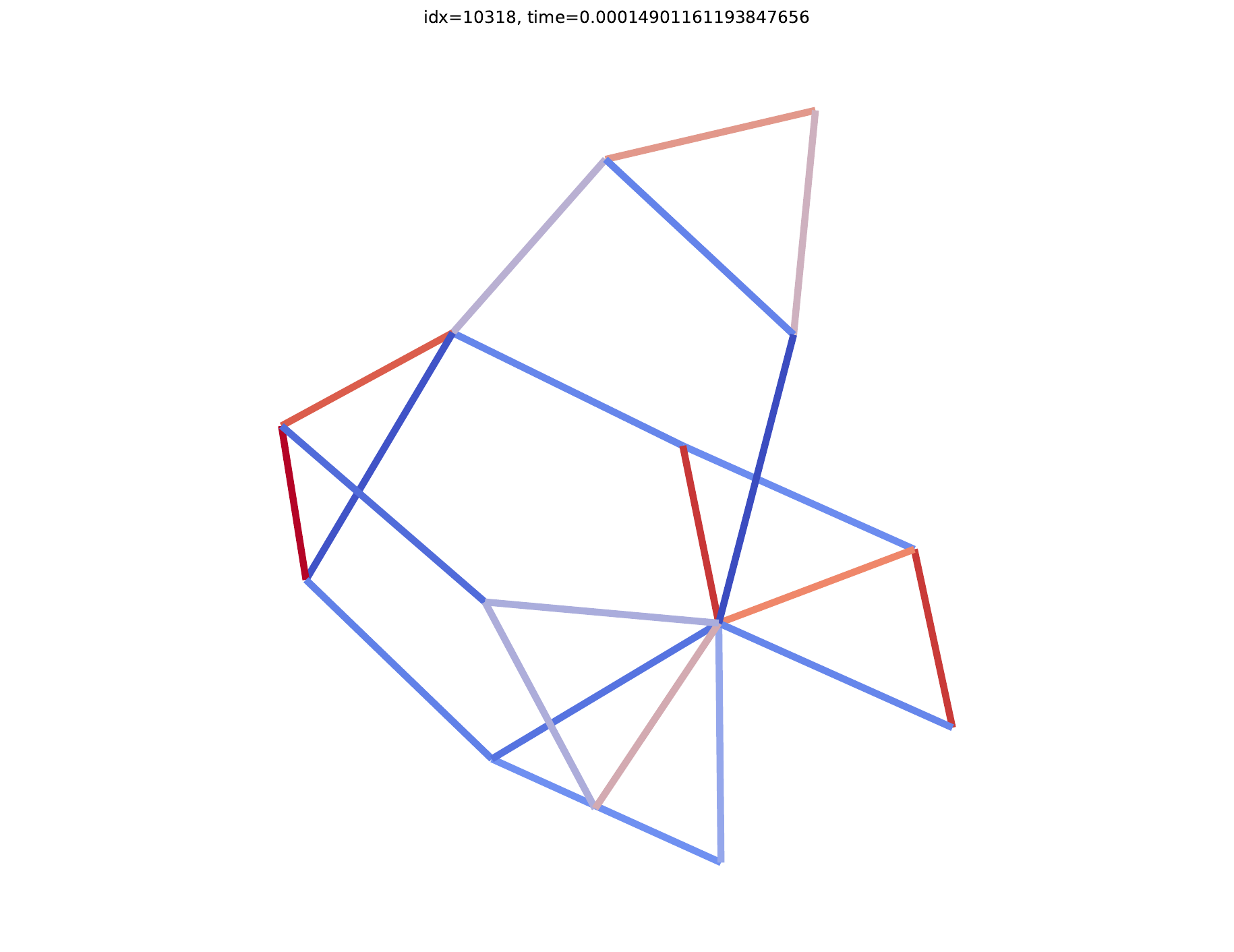} &
\imgcell{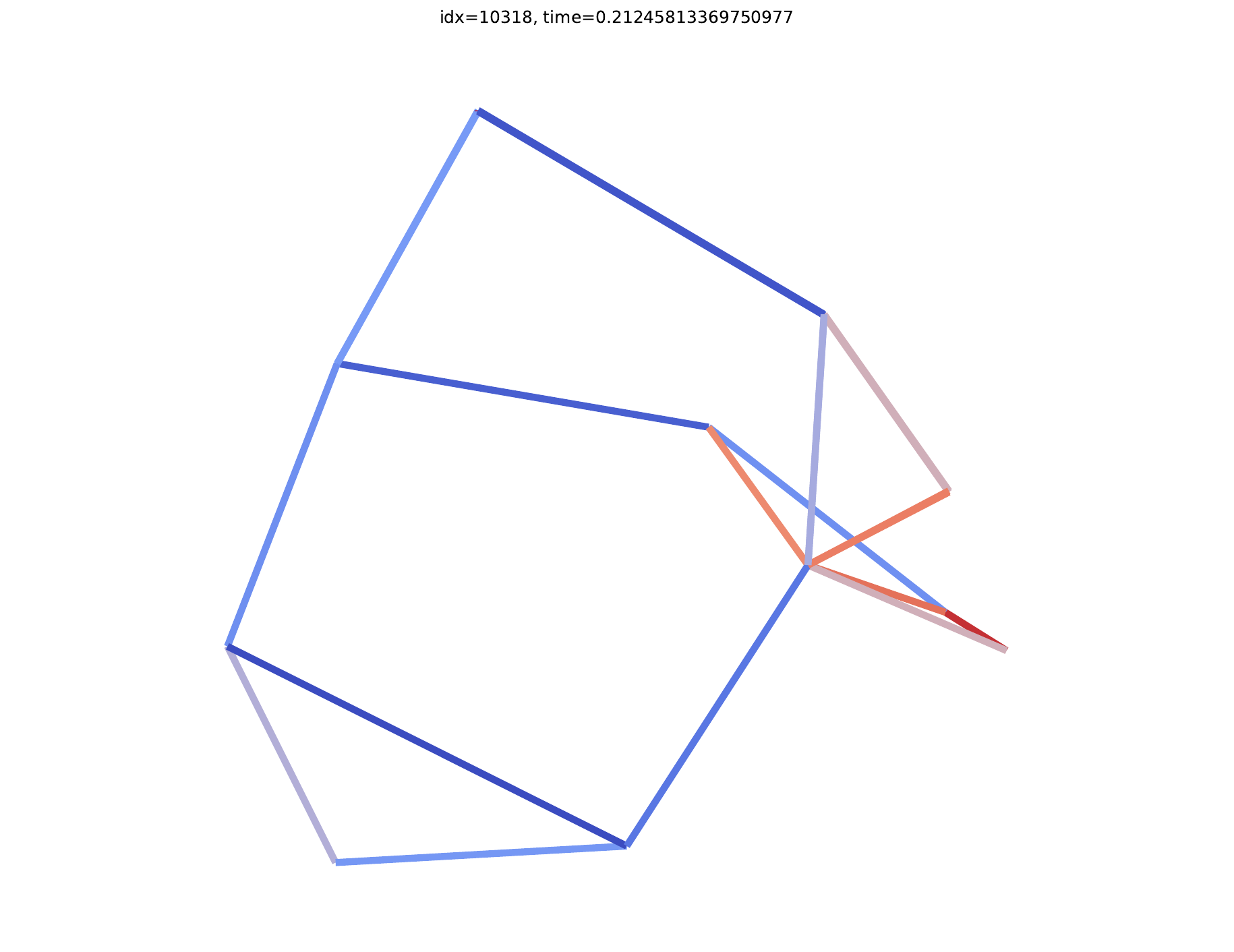} &
\imgcell{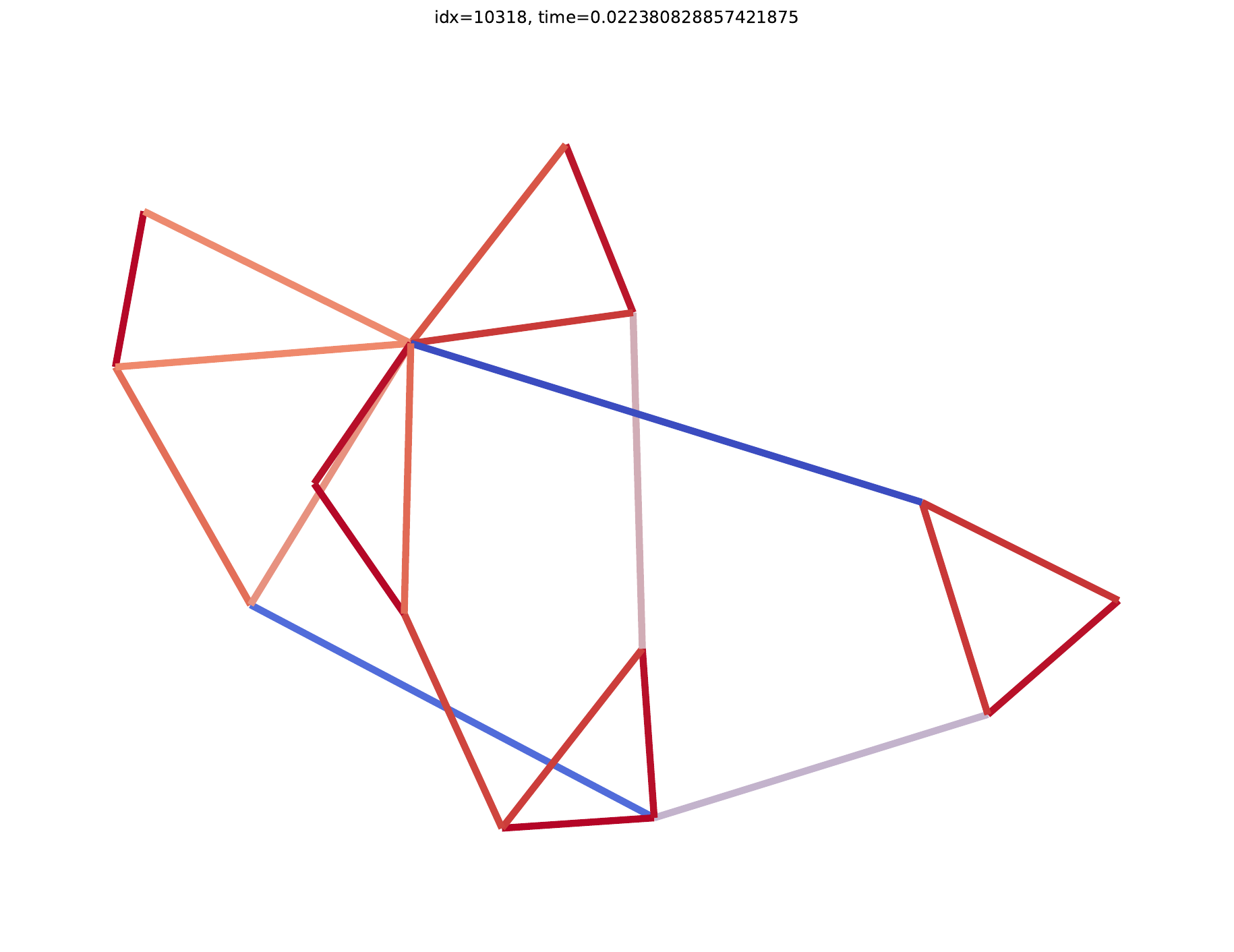} &
\imgcell{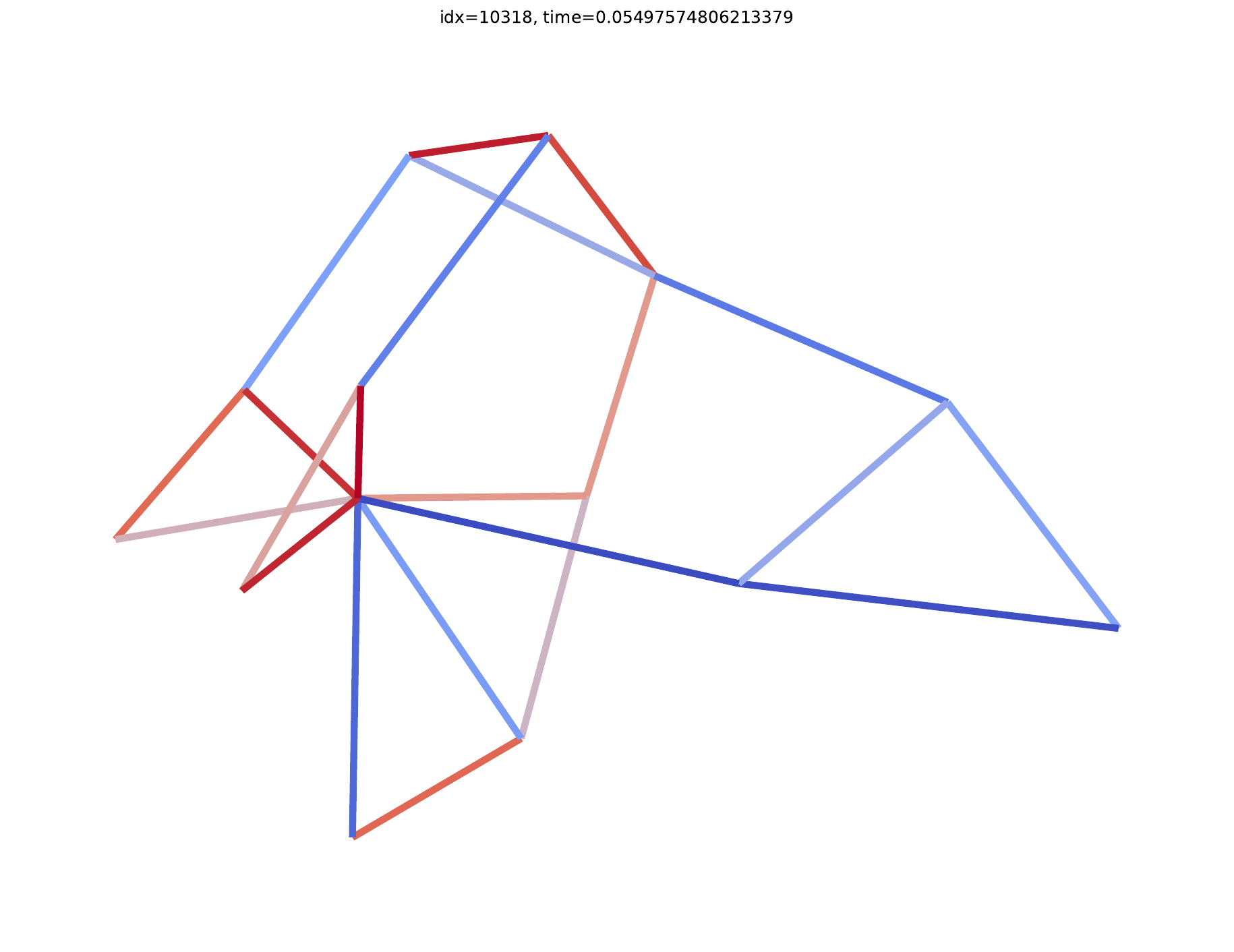} &
\imgcell{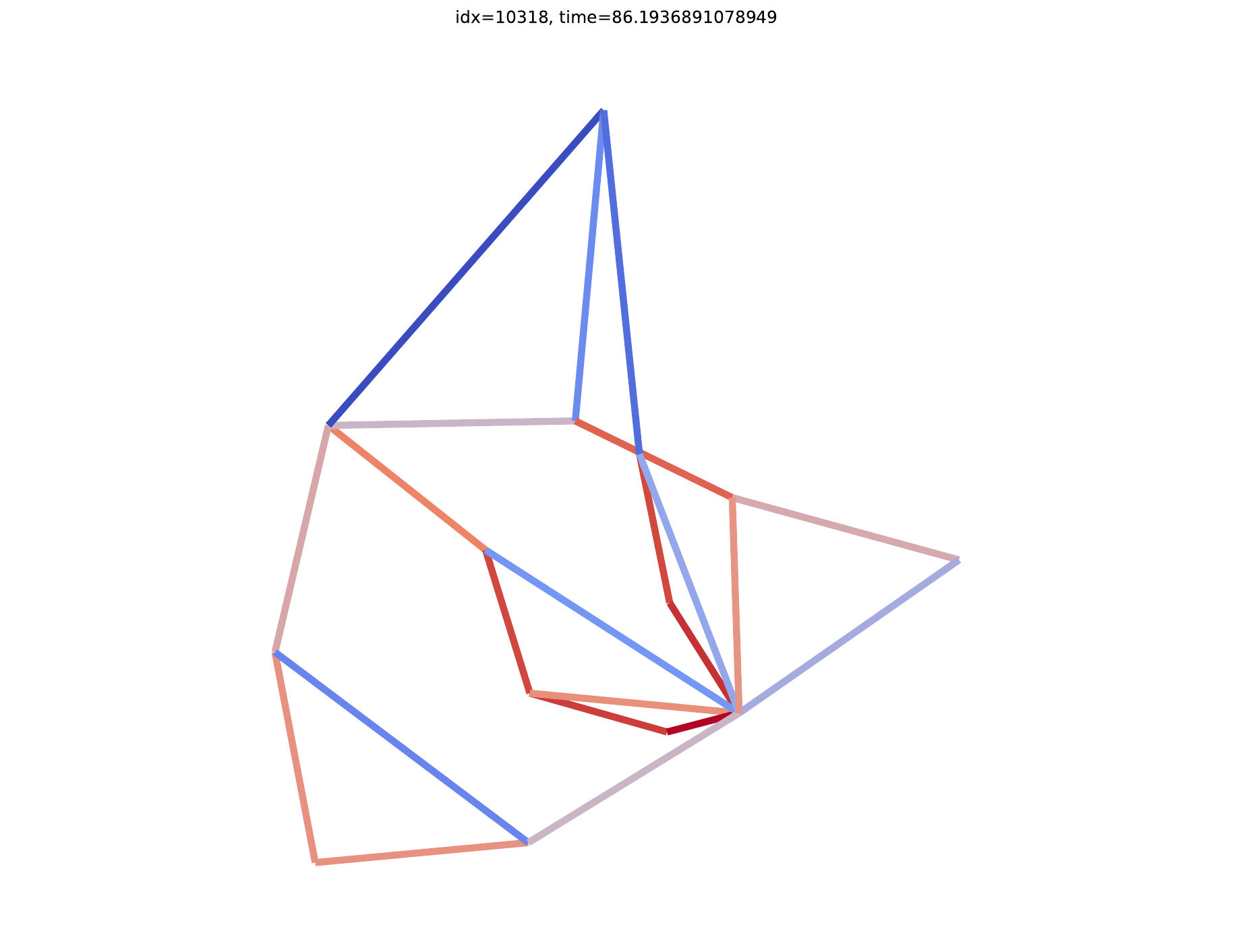} &
\imgcell{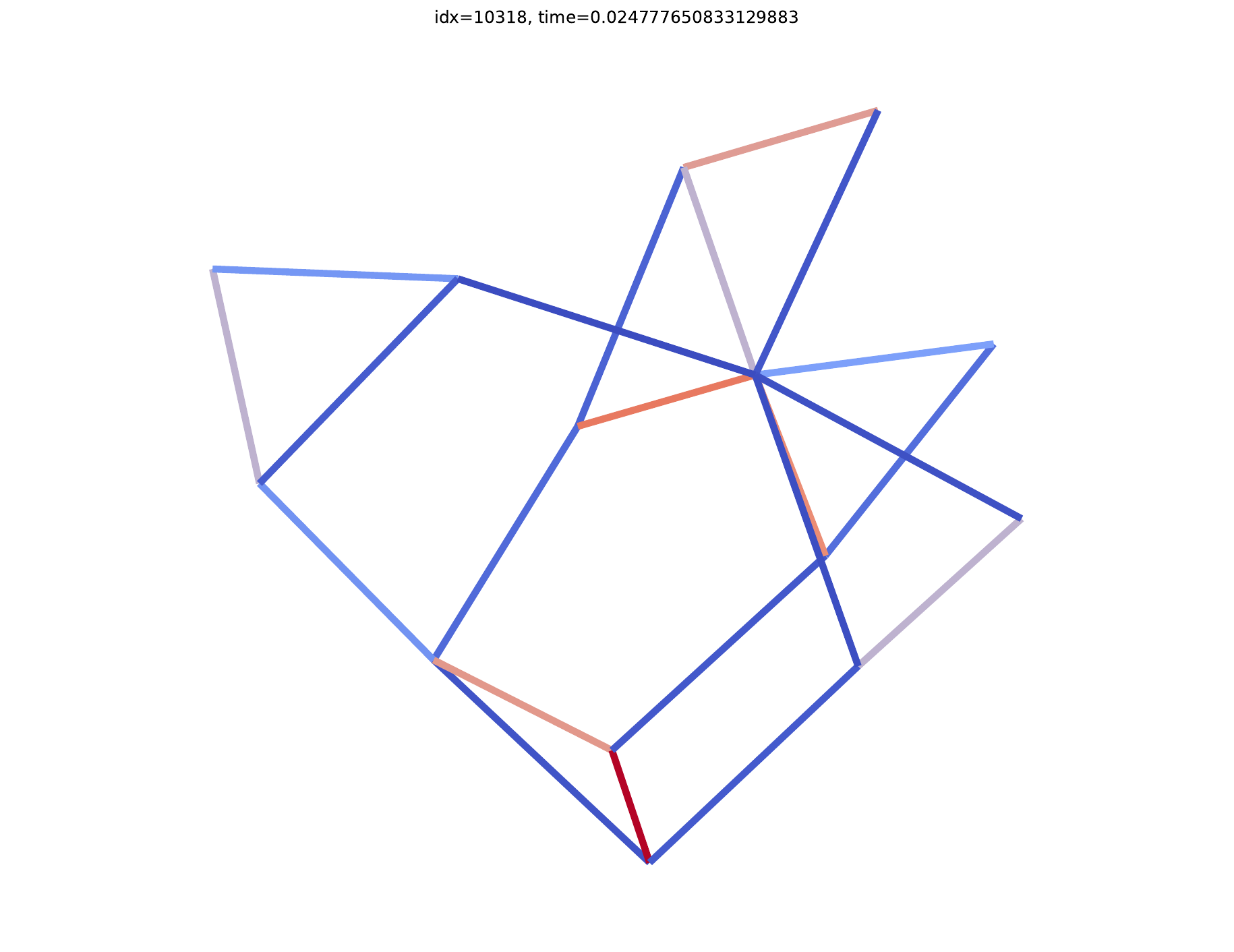} &
\imgcell{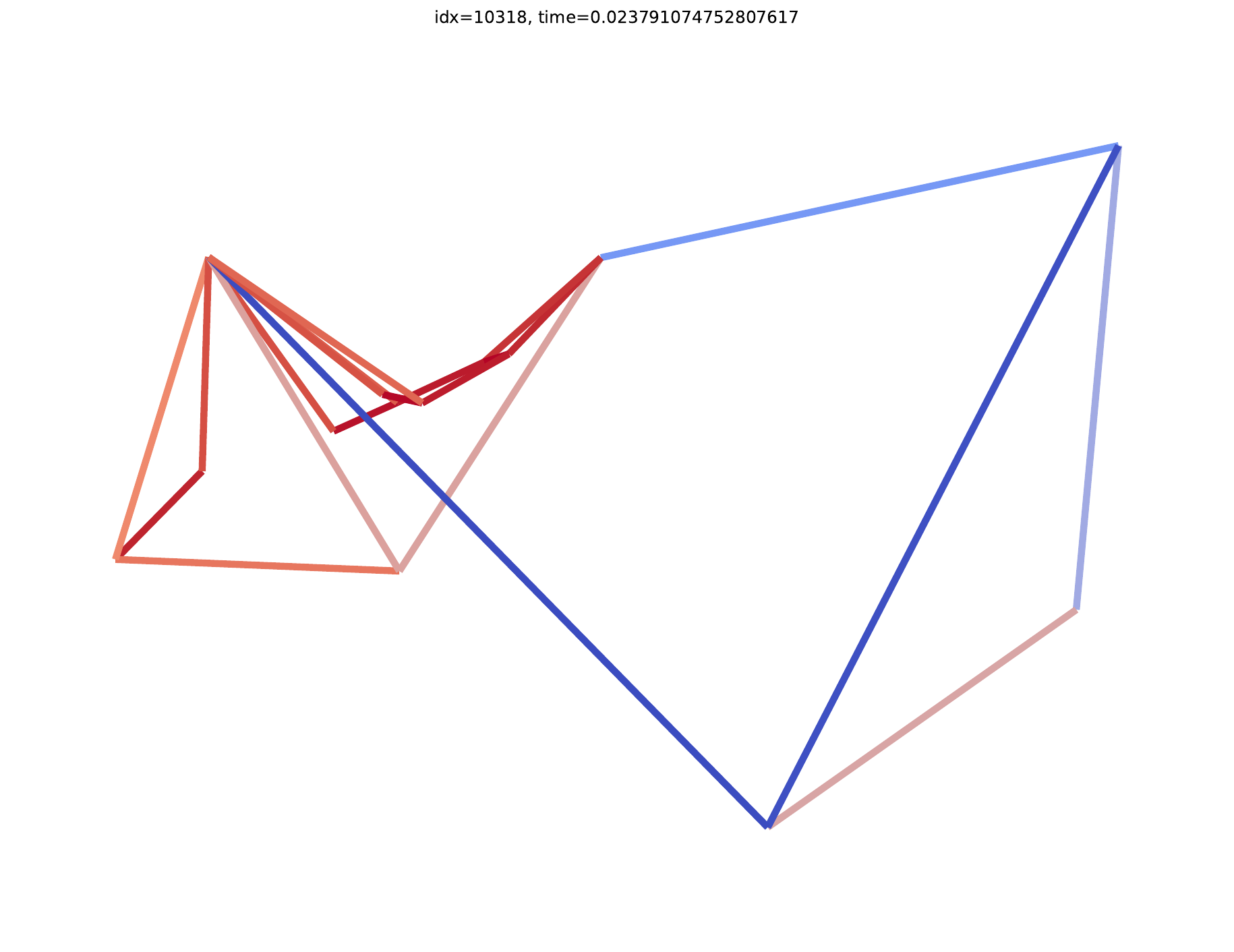} &
\imgcell{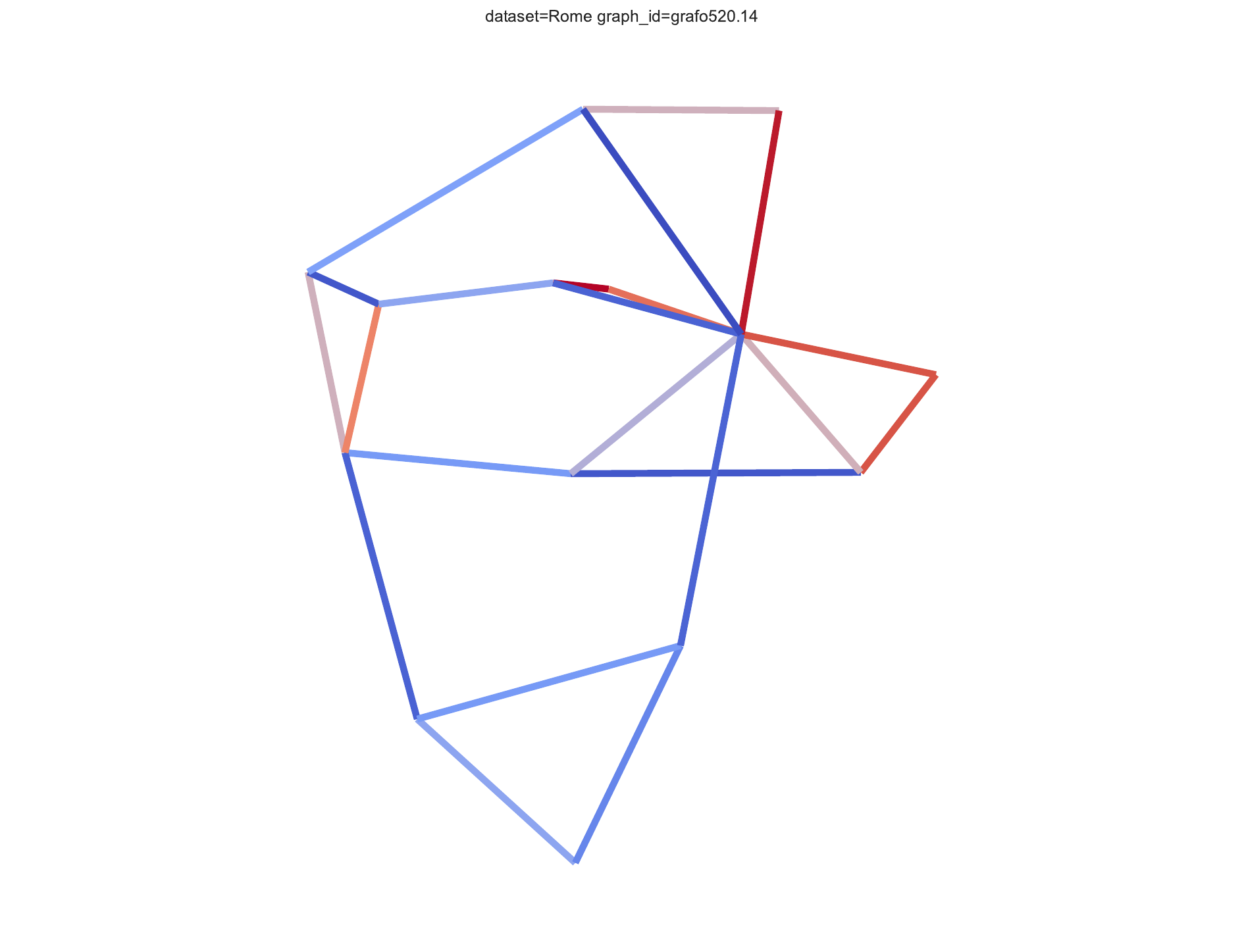} &
\imgcell{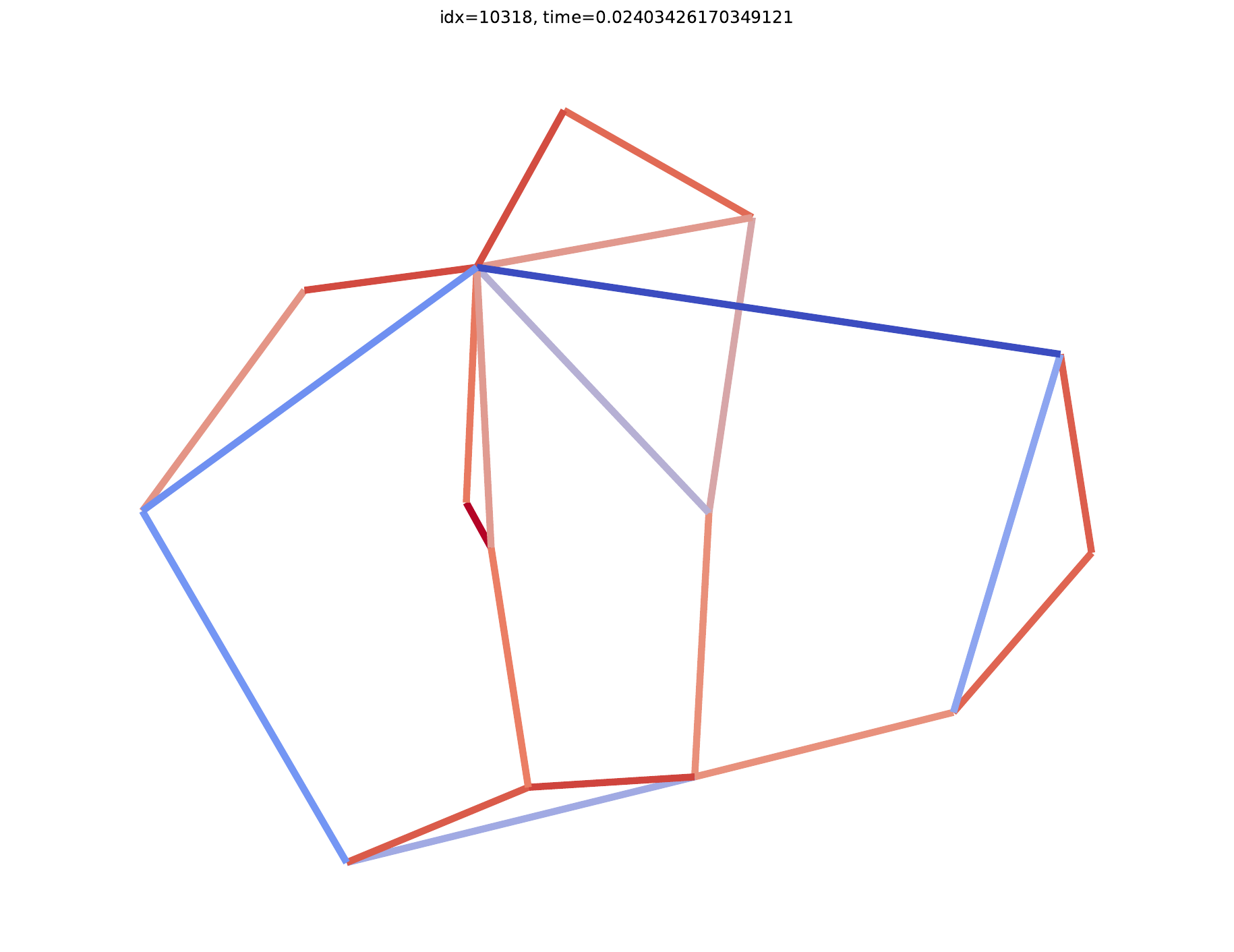} &
\imgcell{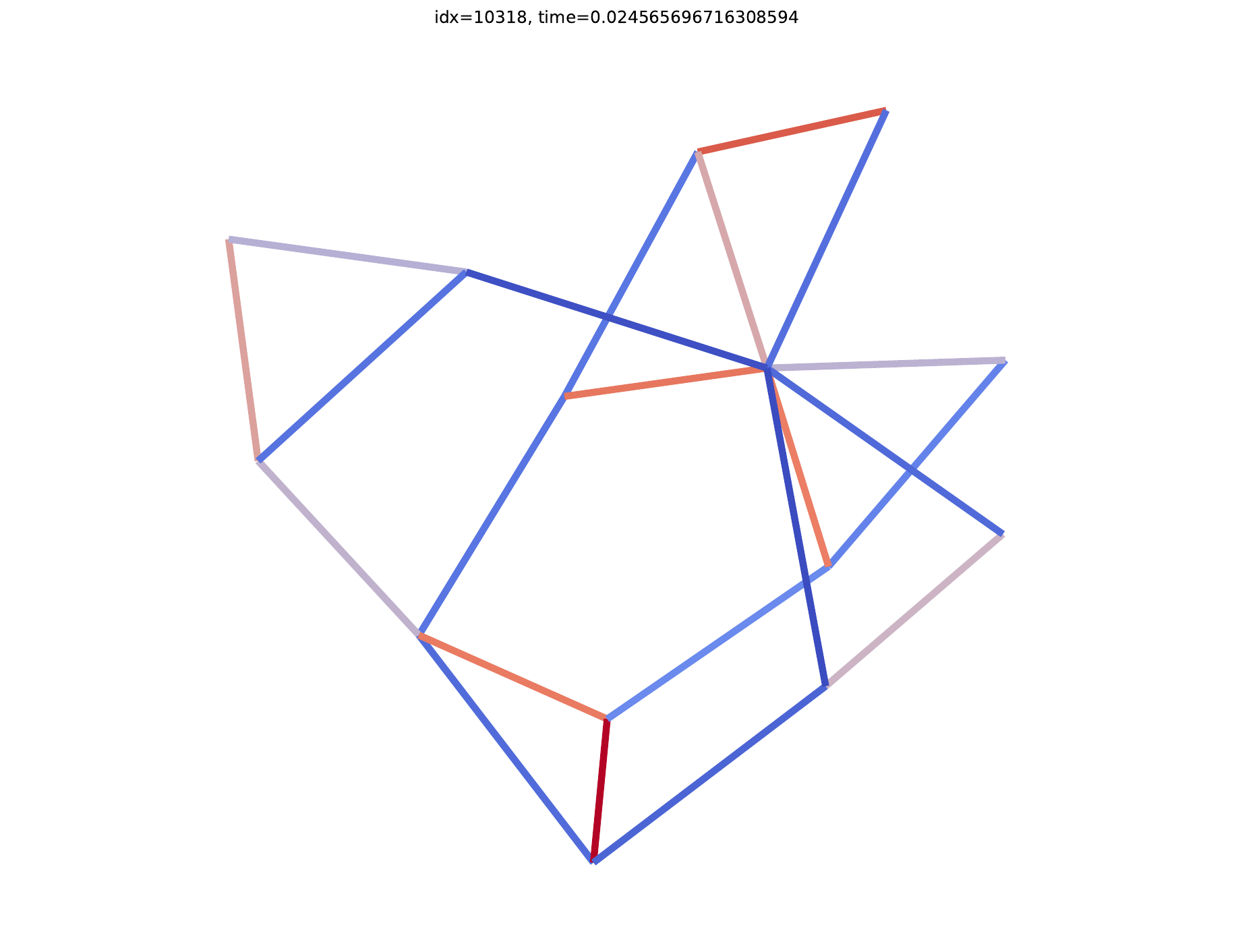} &
\imgcell{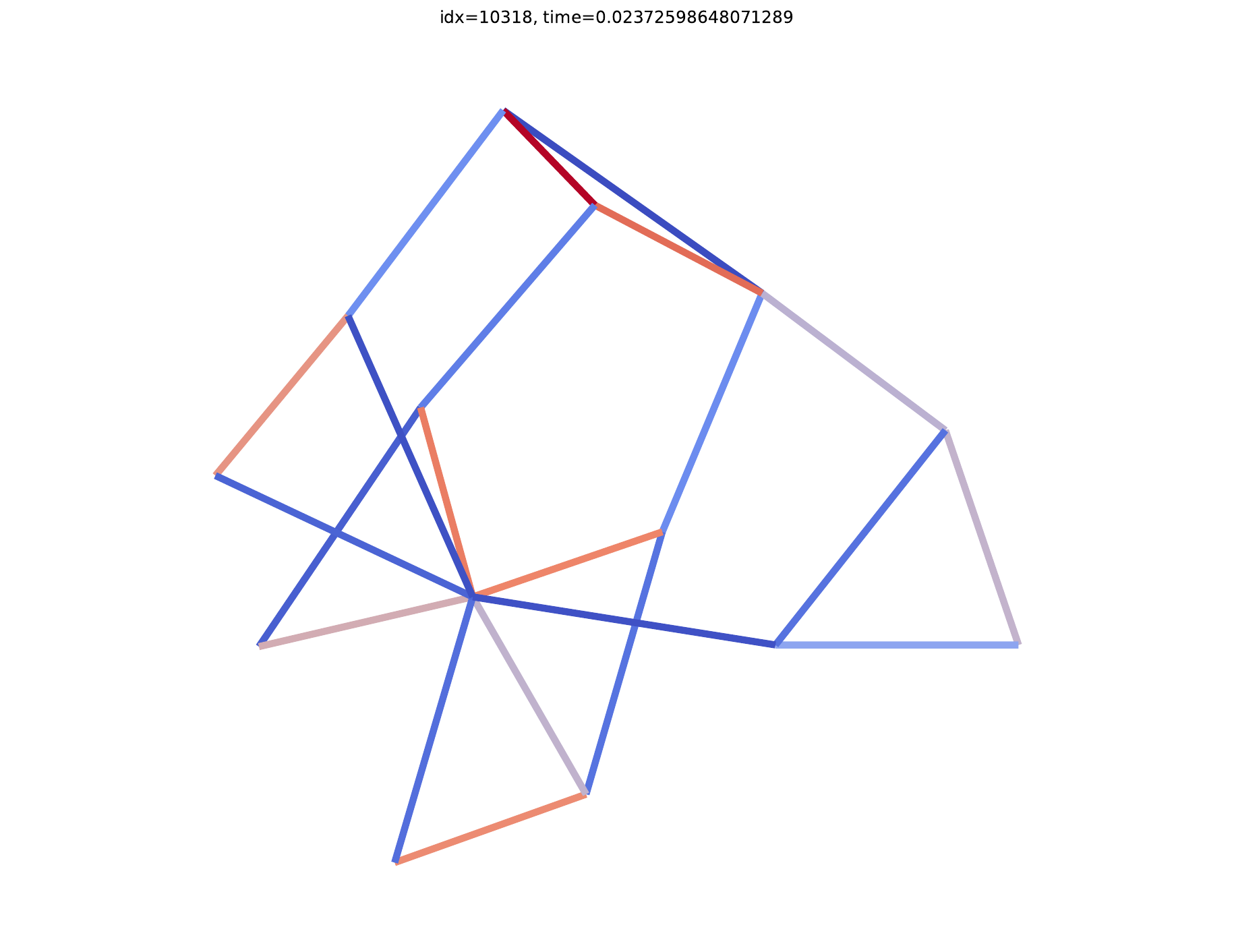} &
\imgcell{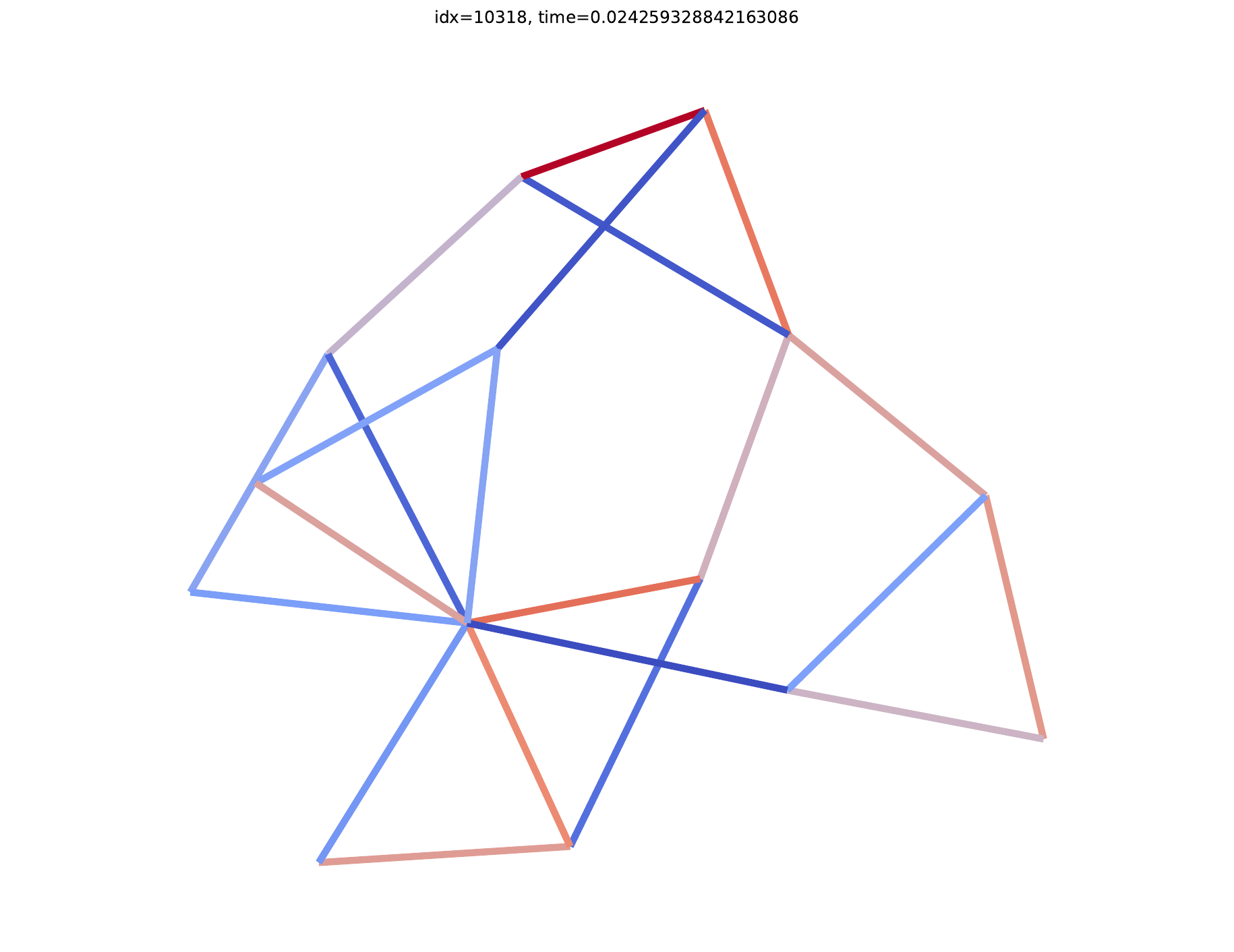} \\

&
t = 0.00s &
t = 0.21s &
t = 0.02s &
t = 0.05s &
t = 86.19s &
t = 0.02s &
t = 0.02s &
t = 0.02s &
t = 0.02s &
t = 0.02s &
t = 0.02s &
t = 0.02s \\

\makecell{\bfseries grafo11328.32\\N = 55\\M = 76} &
\imgcell{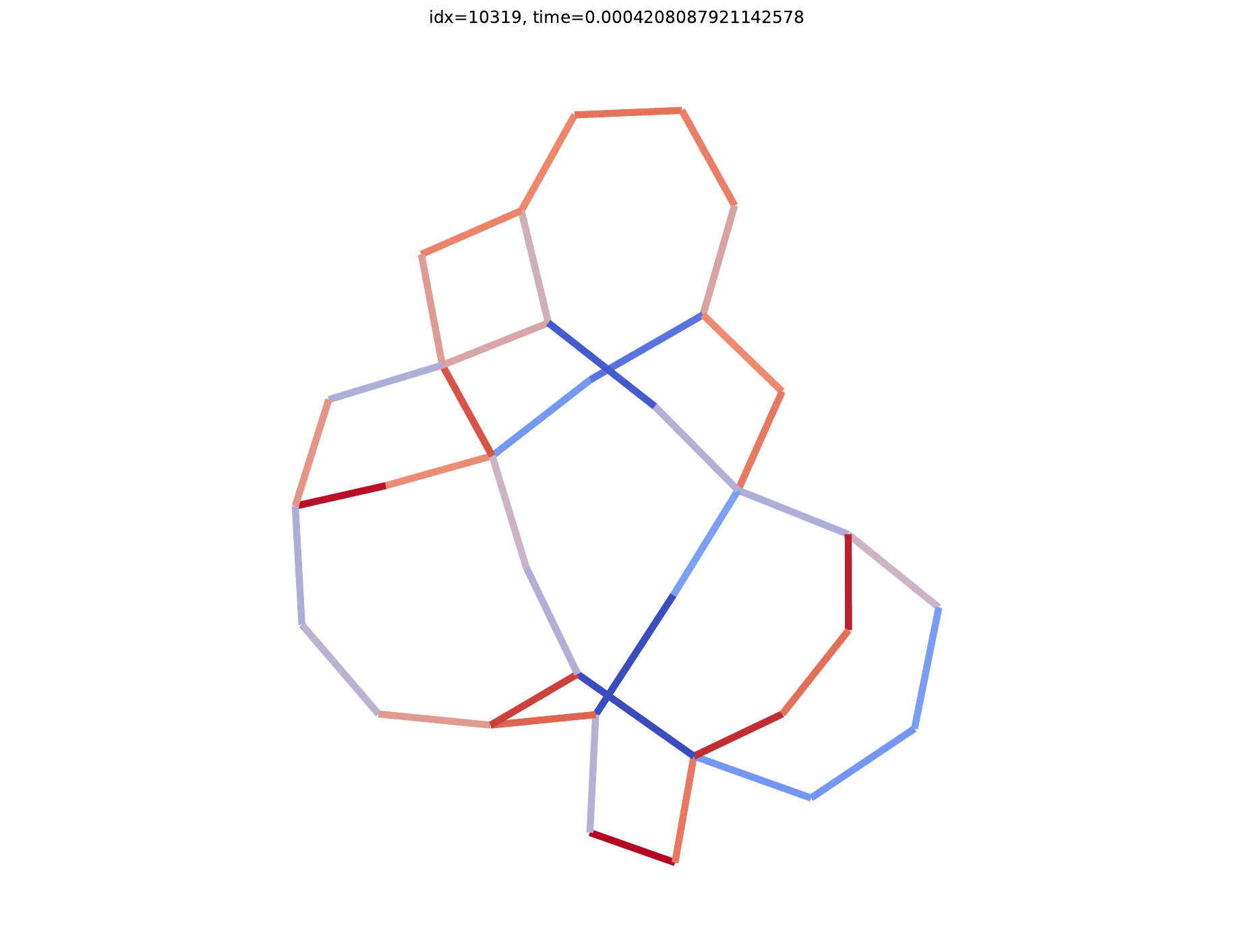} &
\imgcell{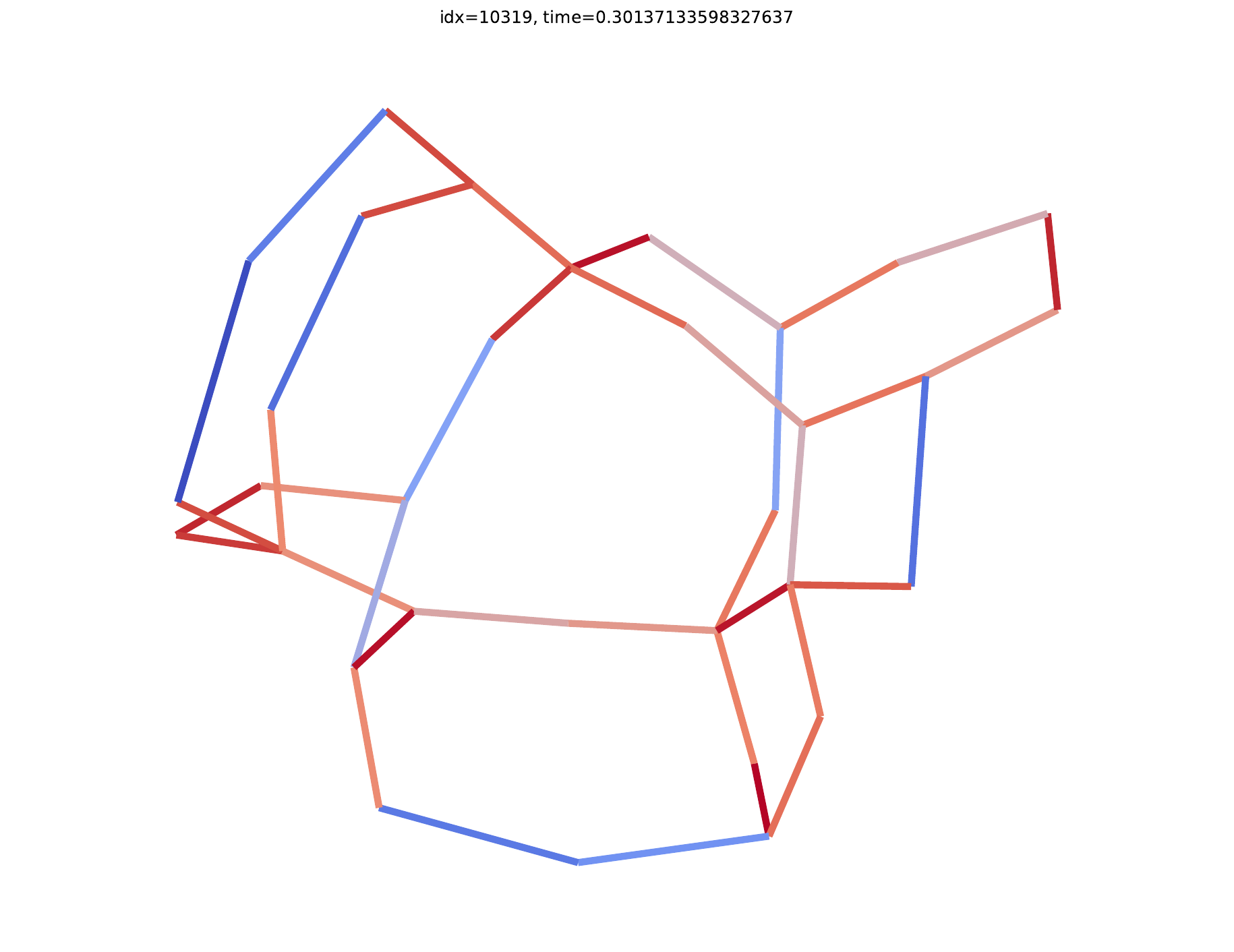} &
\imgcell{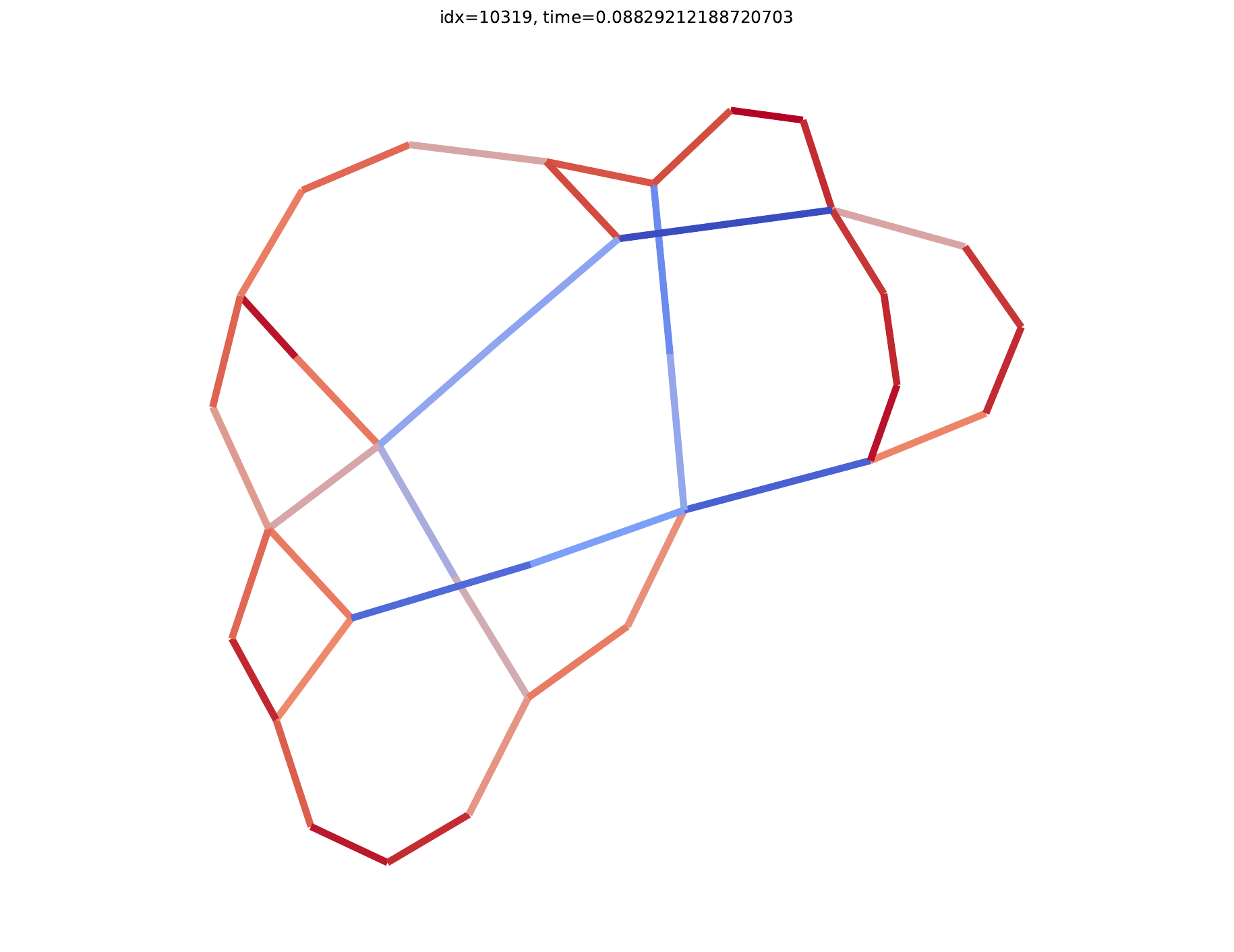} &
\imgcell{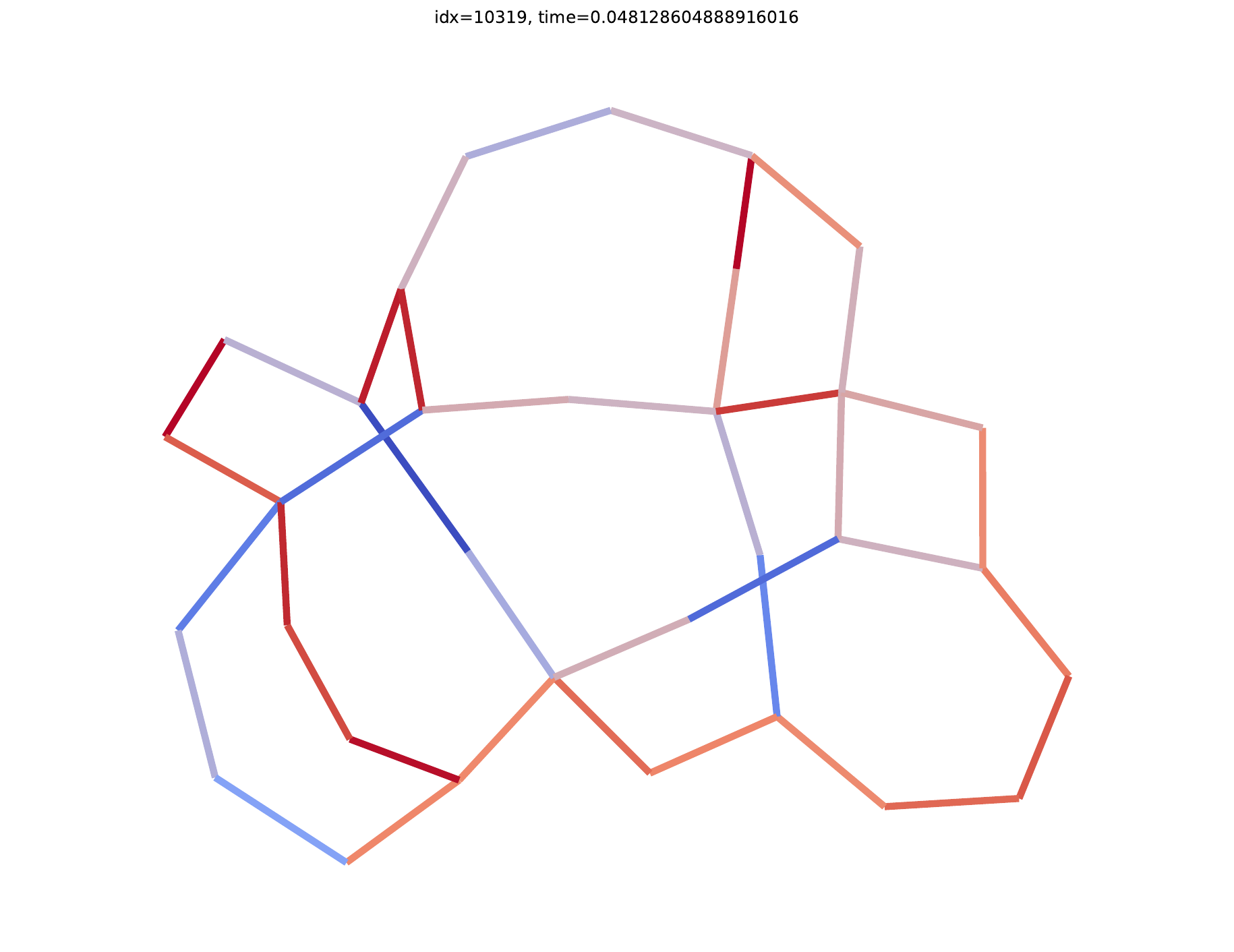} &
\imgcell{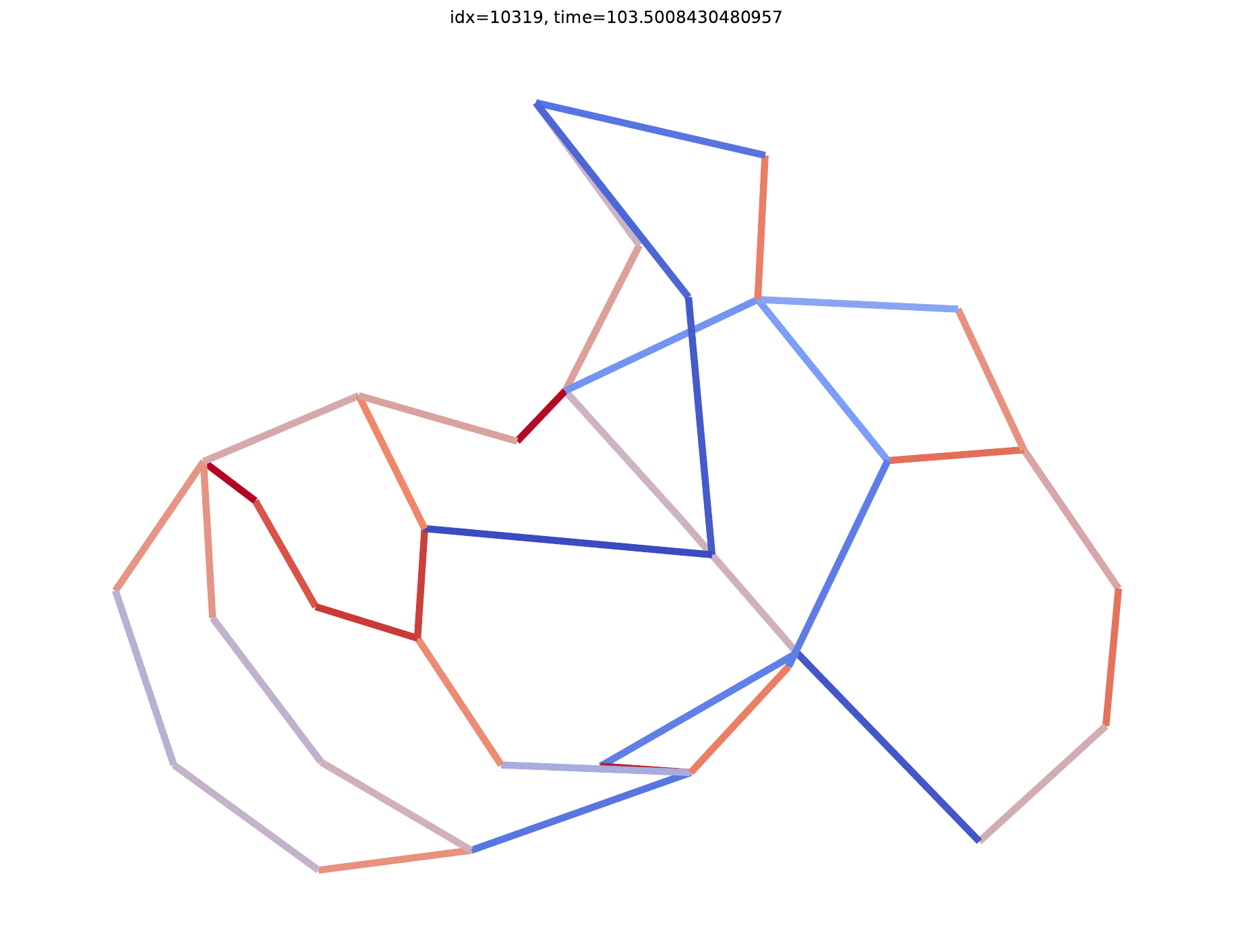} &
\imgcell{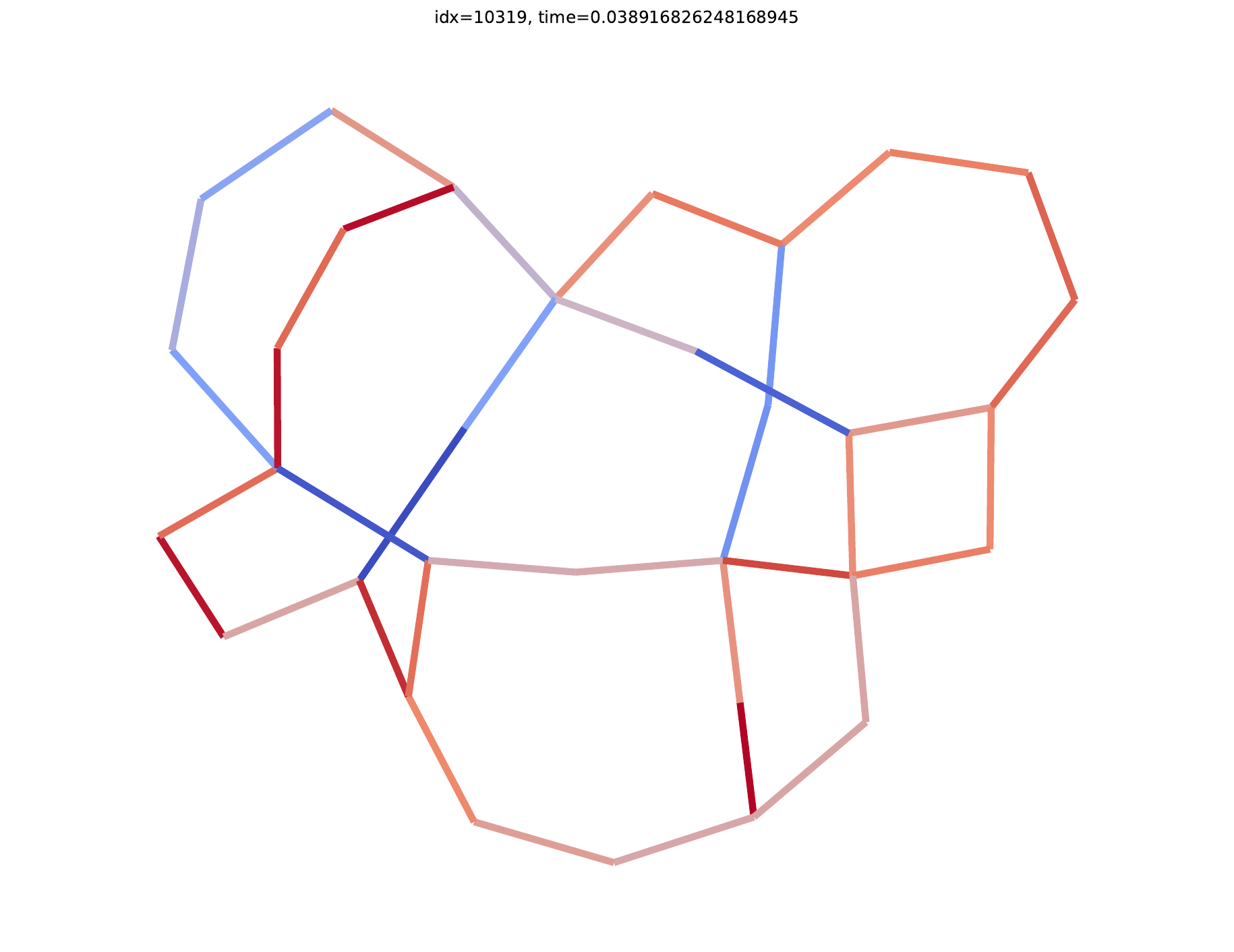} &
\imgcell{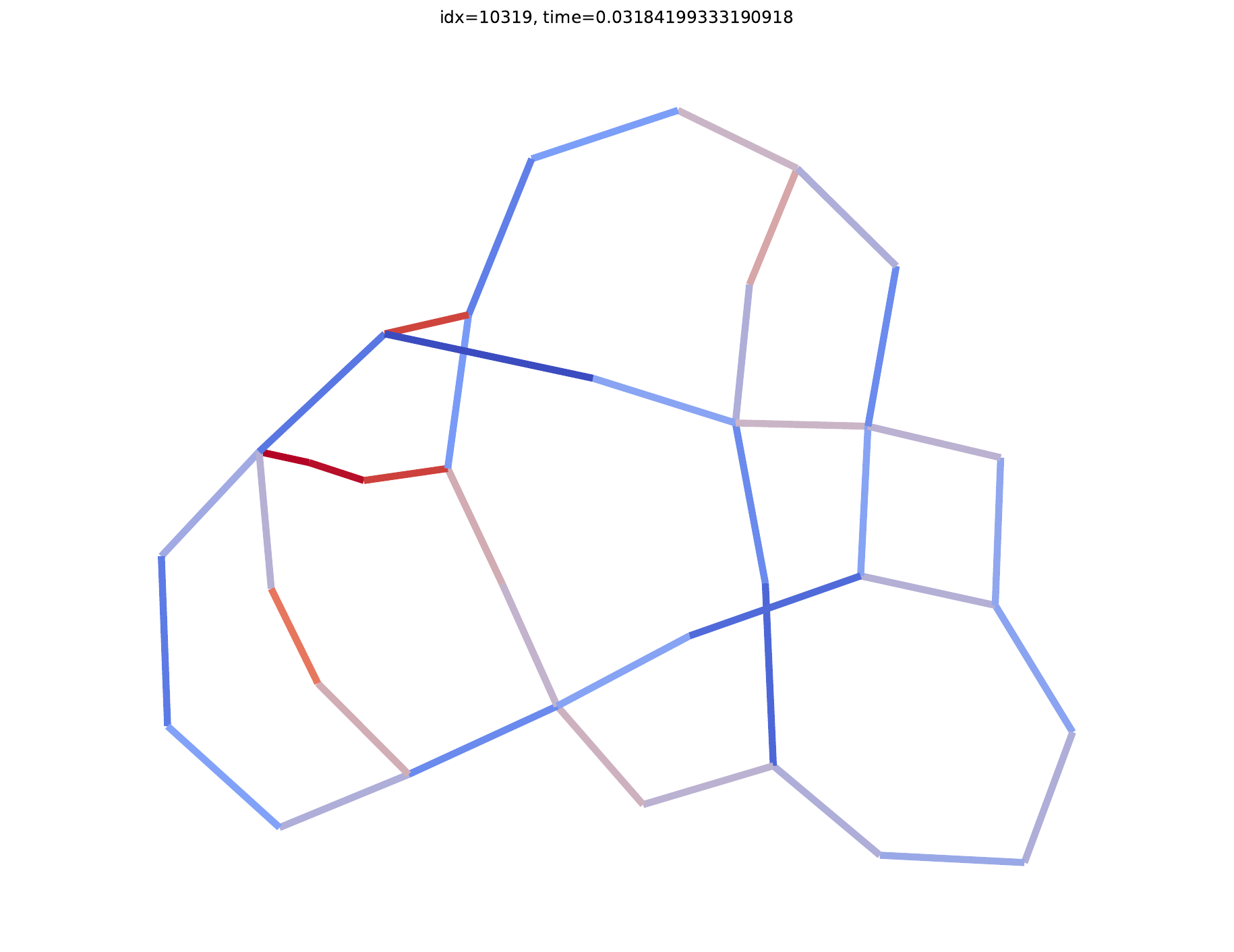} &
\imgcell{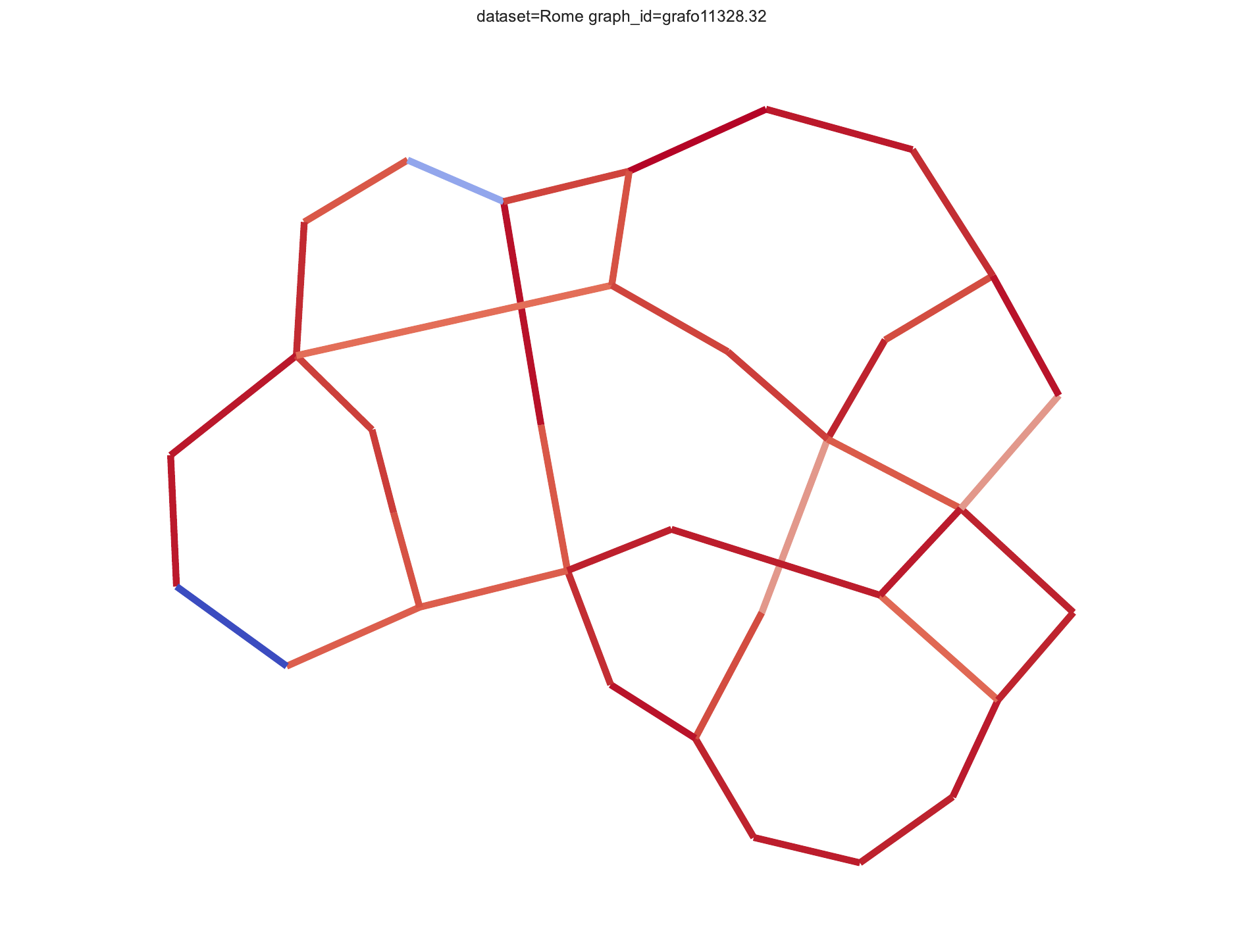} &
\imgcell{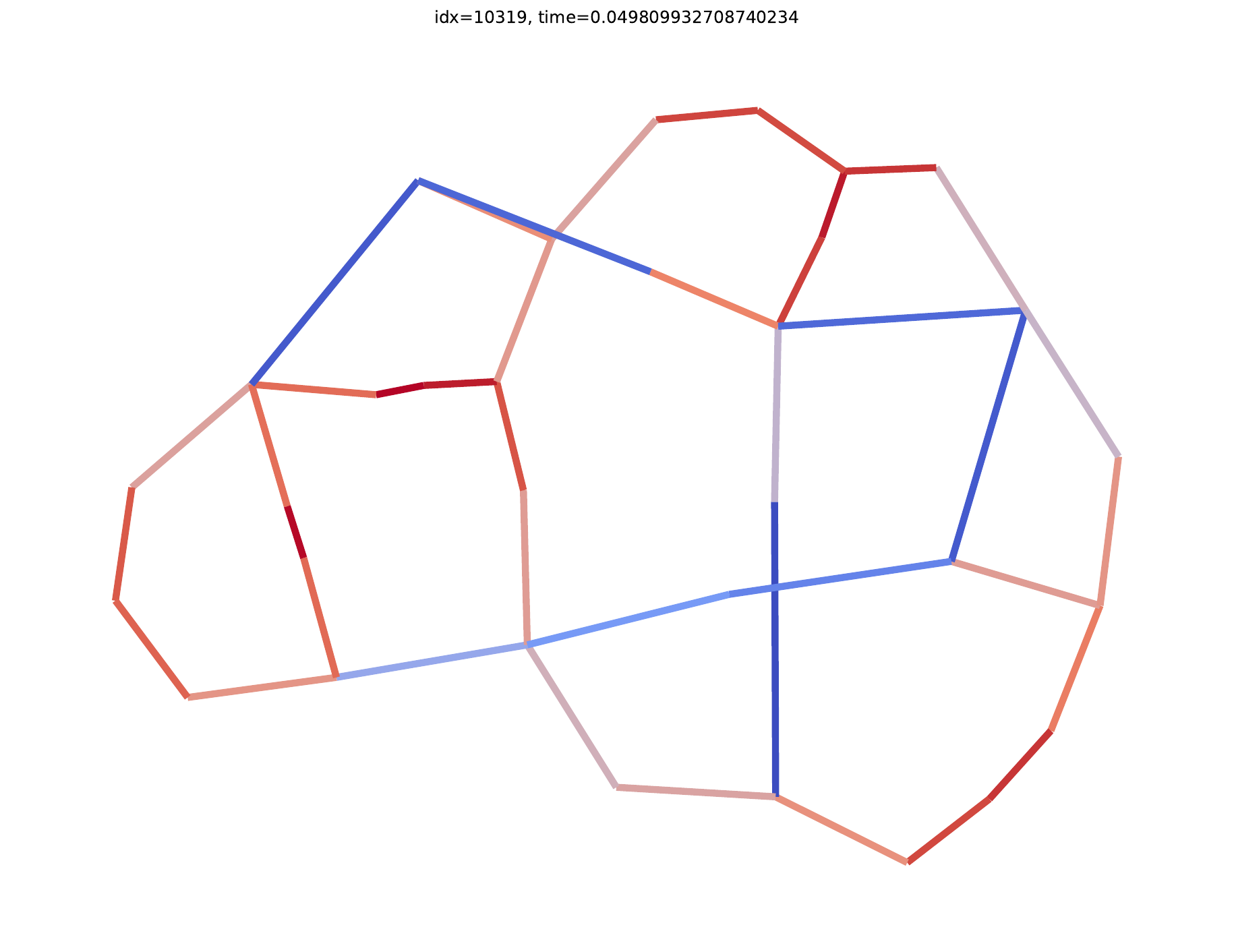} &
\imgcell{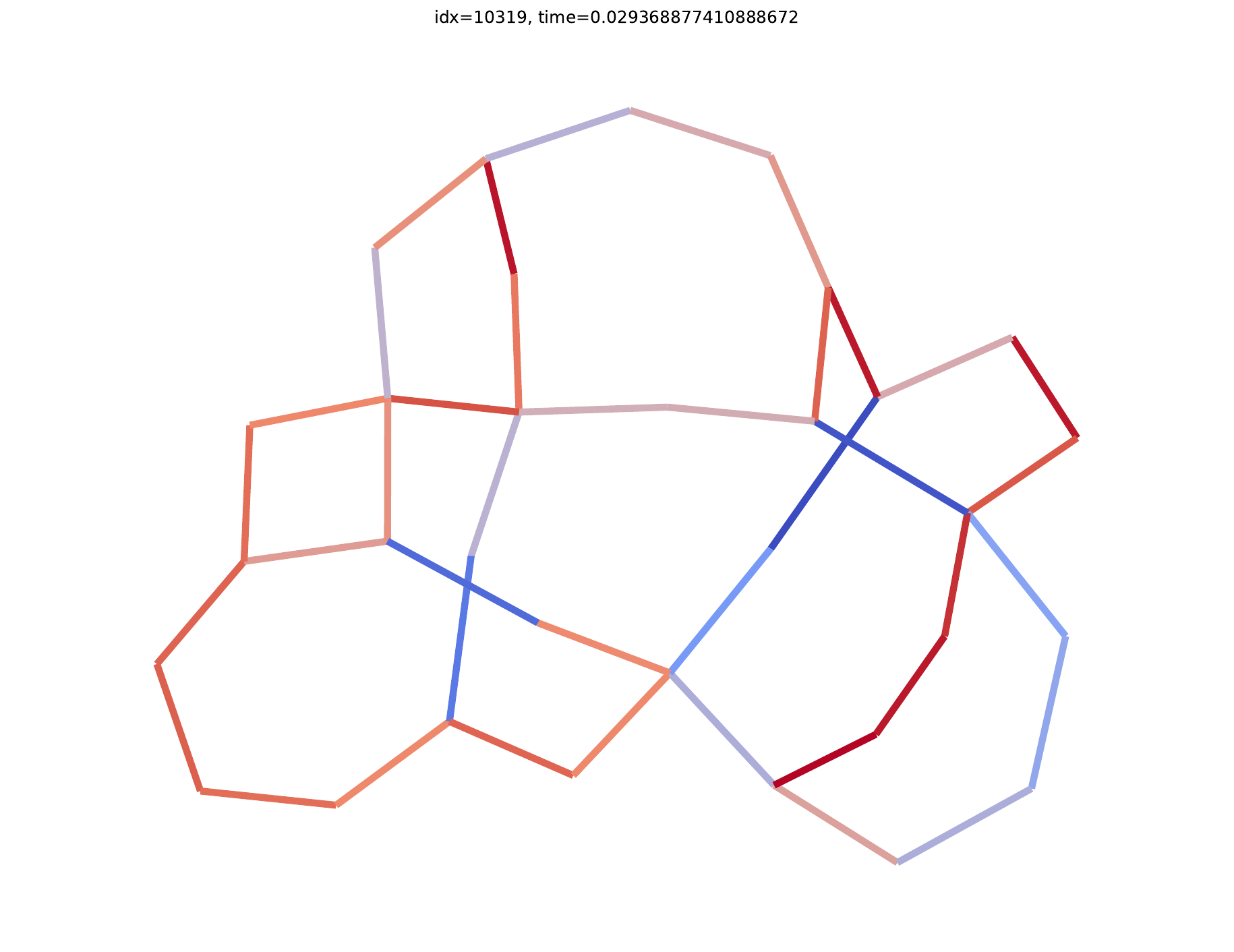} &
\imgcell{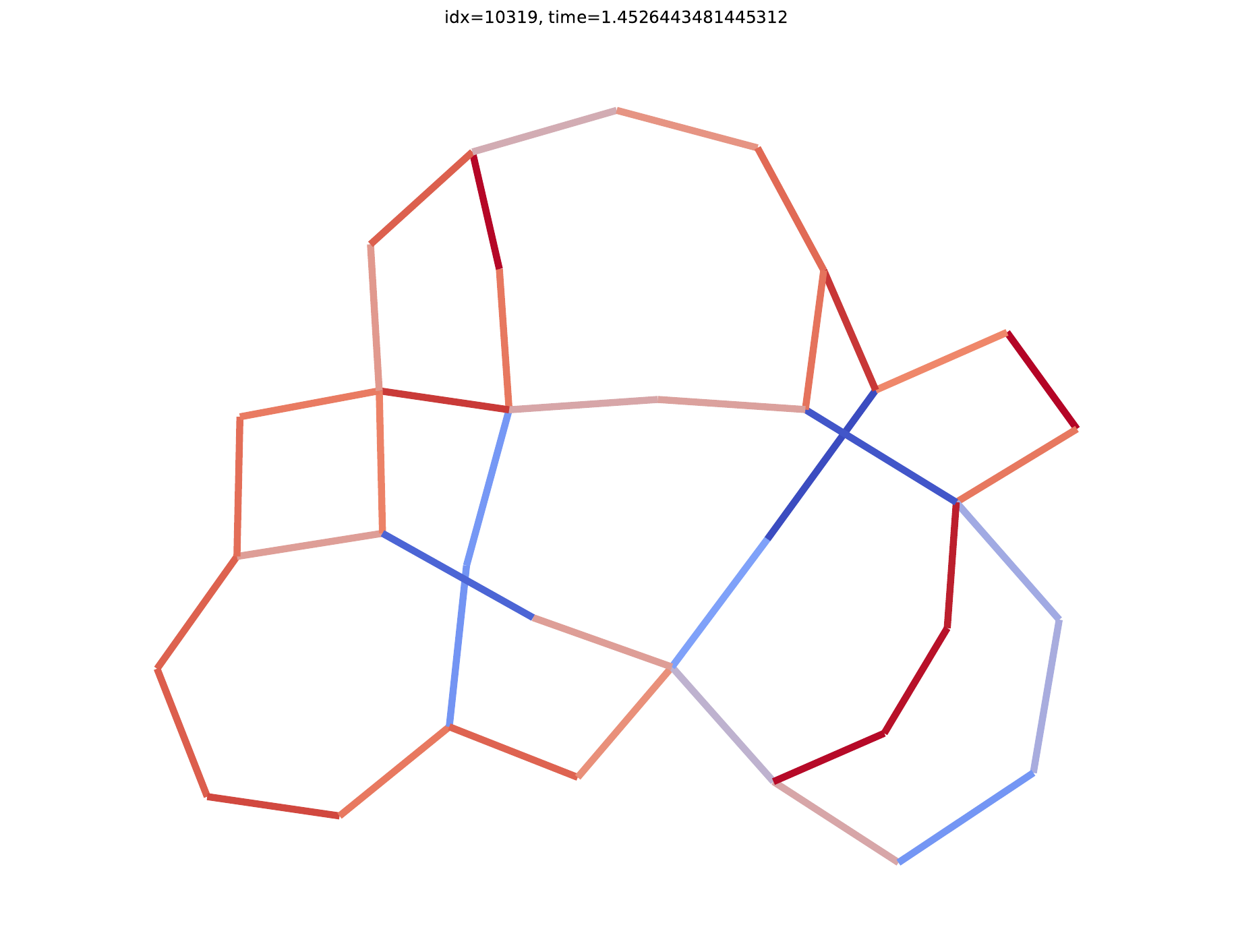} &
\imgcell{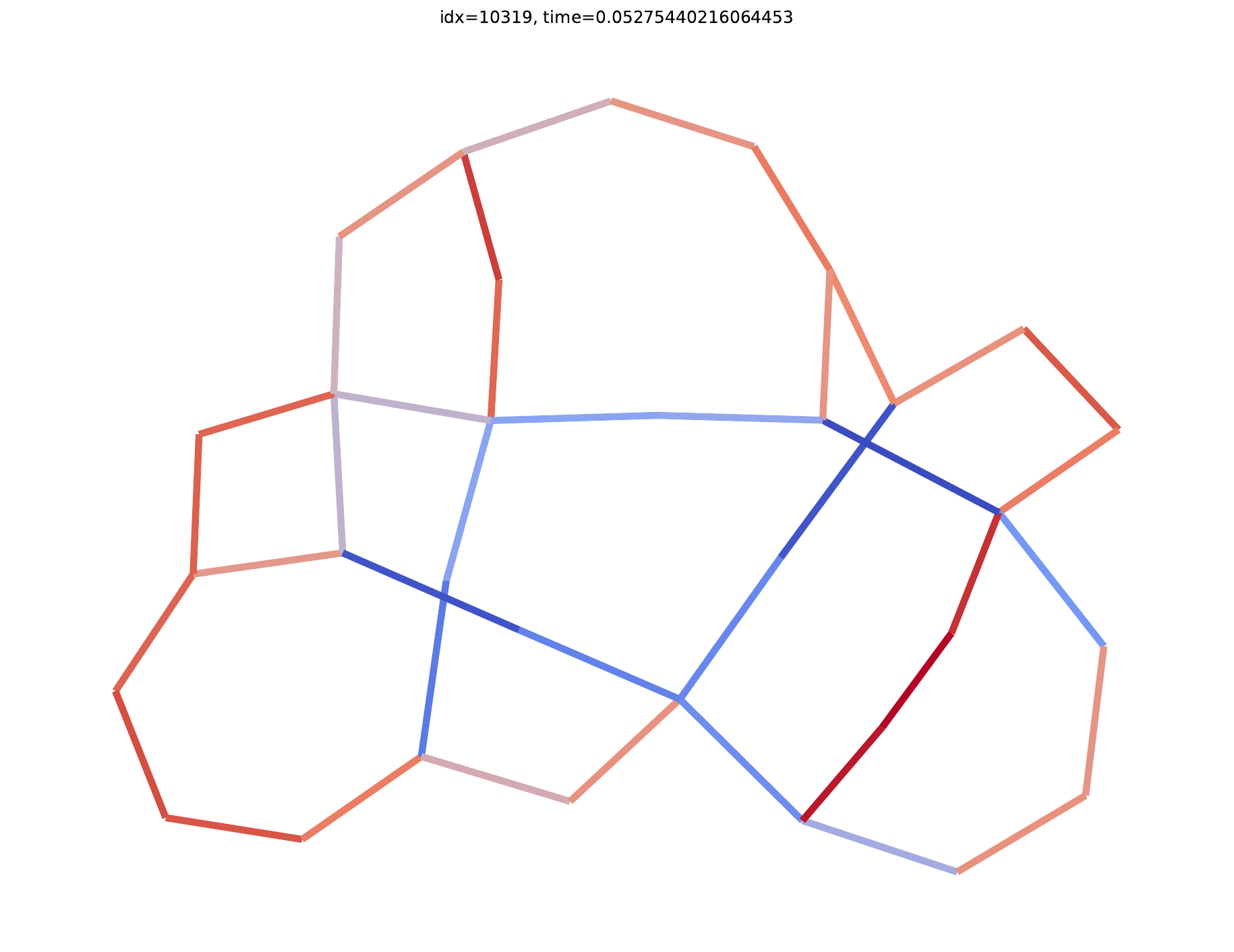} \\

&
t = 0.00s &
t = 0.30s &
t = 0.09s &
t = 0.05s &
t = 103.50s &
t = 0.04s &
t = 0.03s &
t = 0.04s &
t = 0.05s &
t = 0.03s &
t = 0.05s &
t = 0.05s \\

\makecell{\bfseries grafo5584.48\\N = 82\\M = 113} &
\imgcell{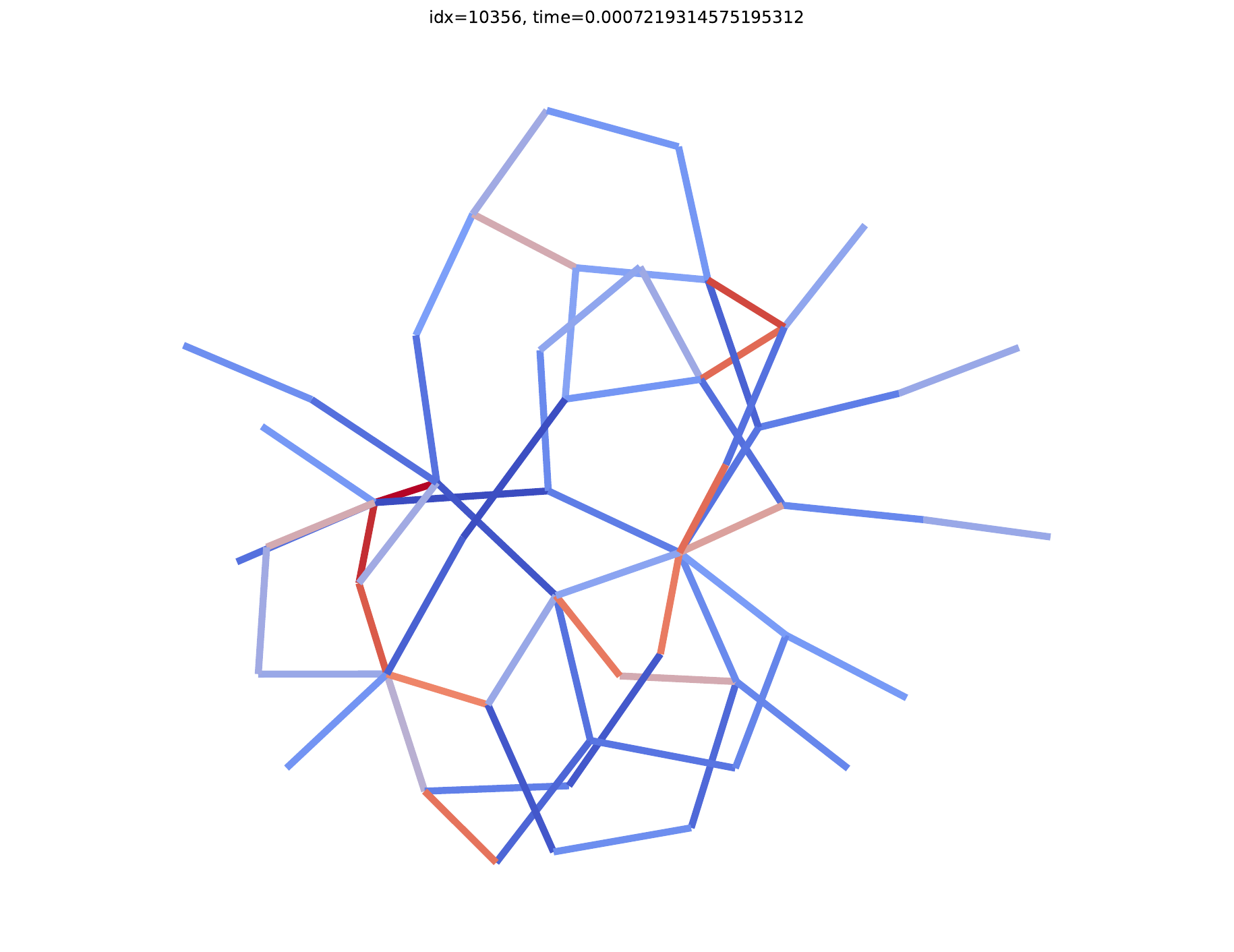} &
\imgcell{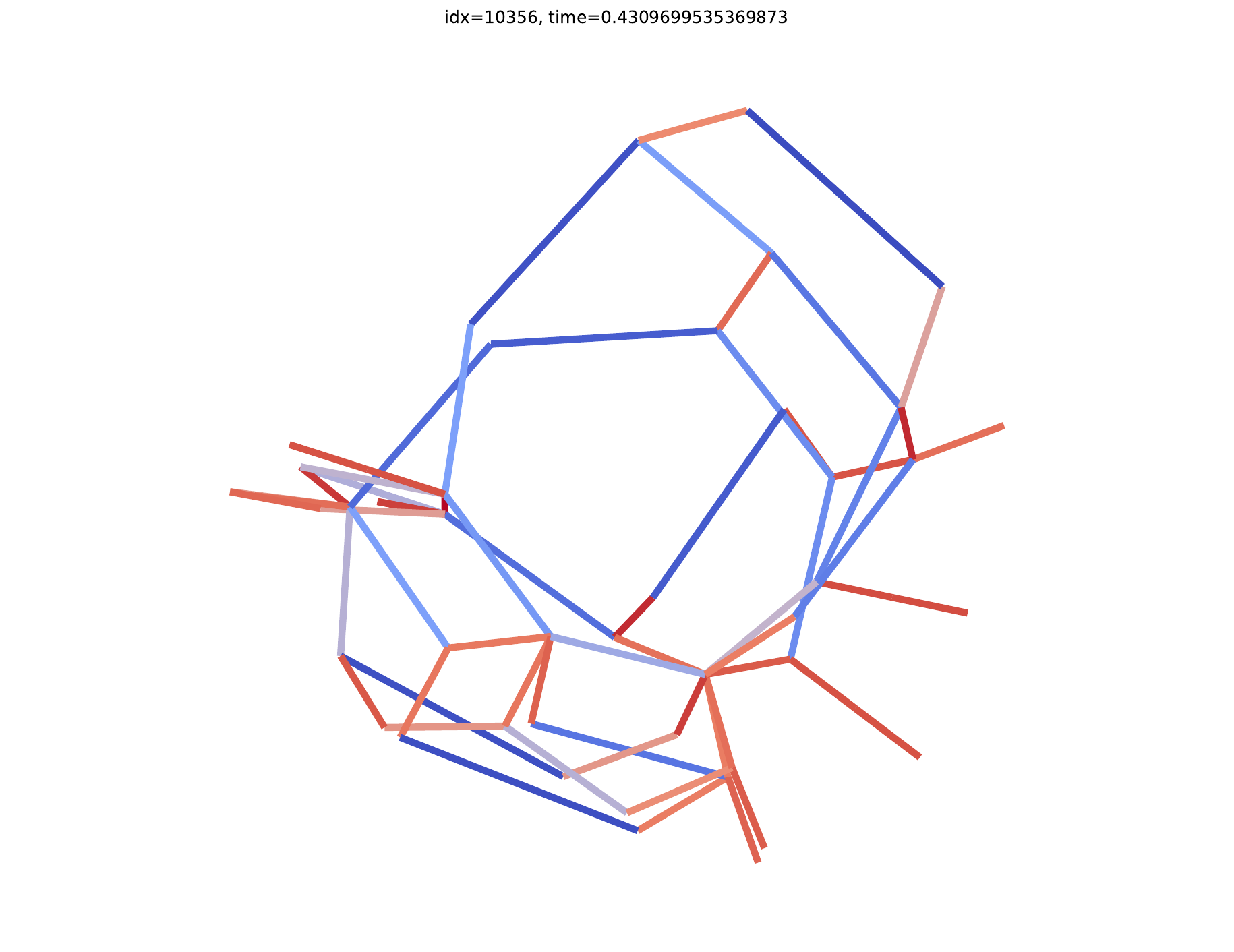} &
\imgcell{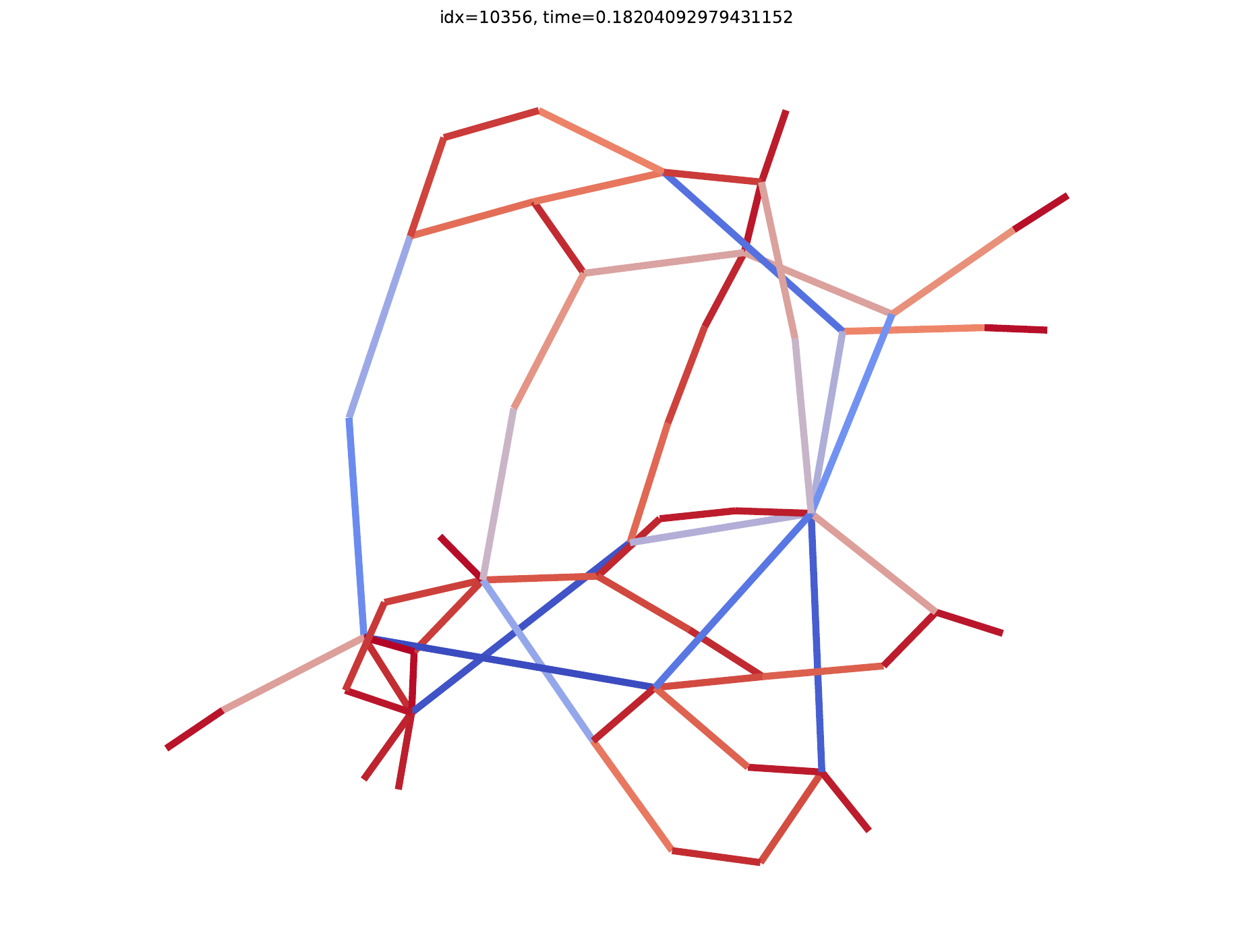} &
\imgcell{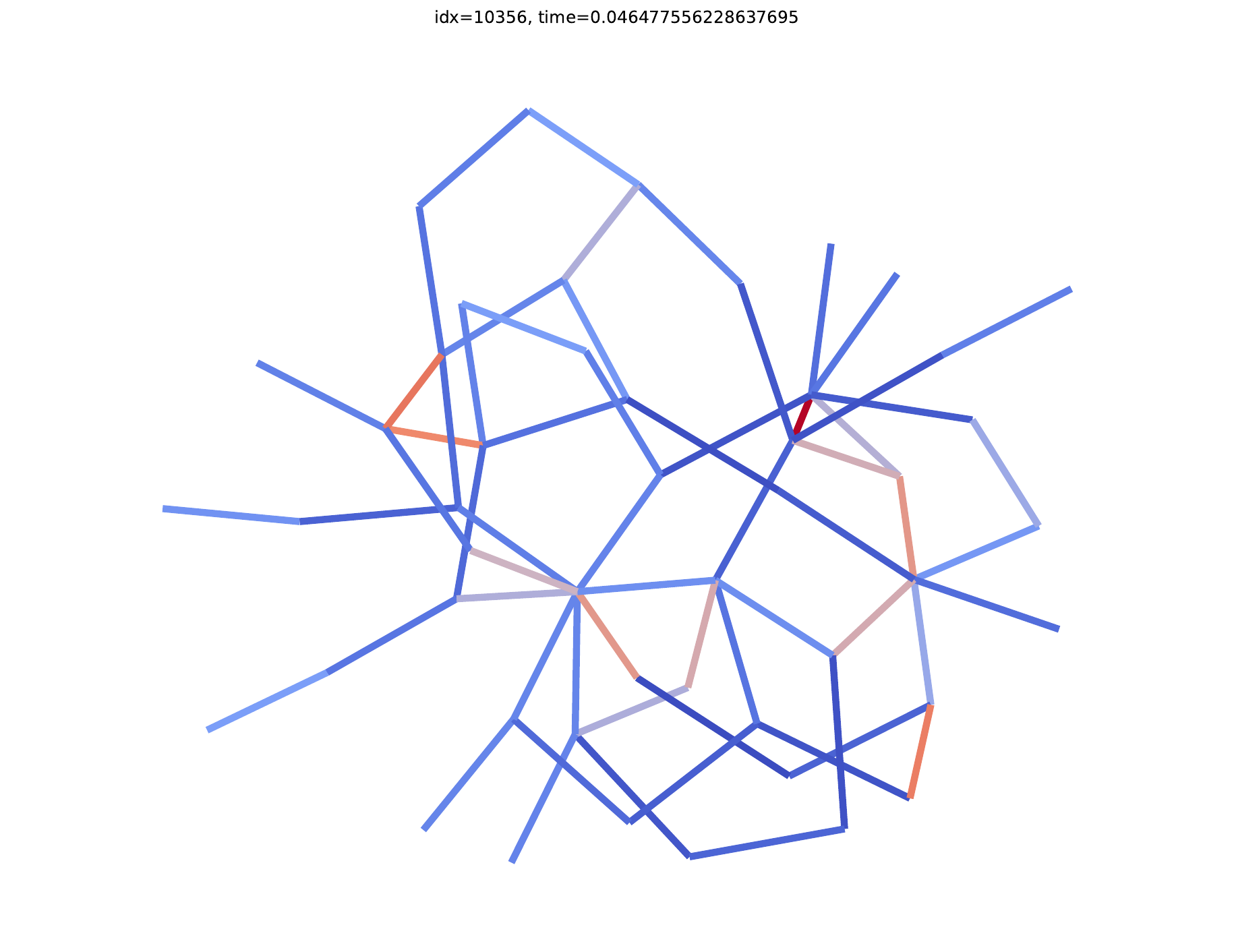} &
\imgcell{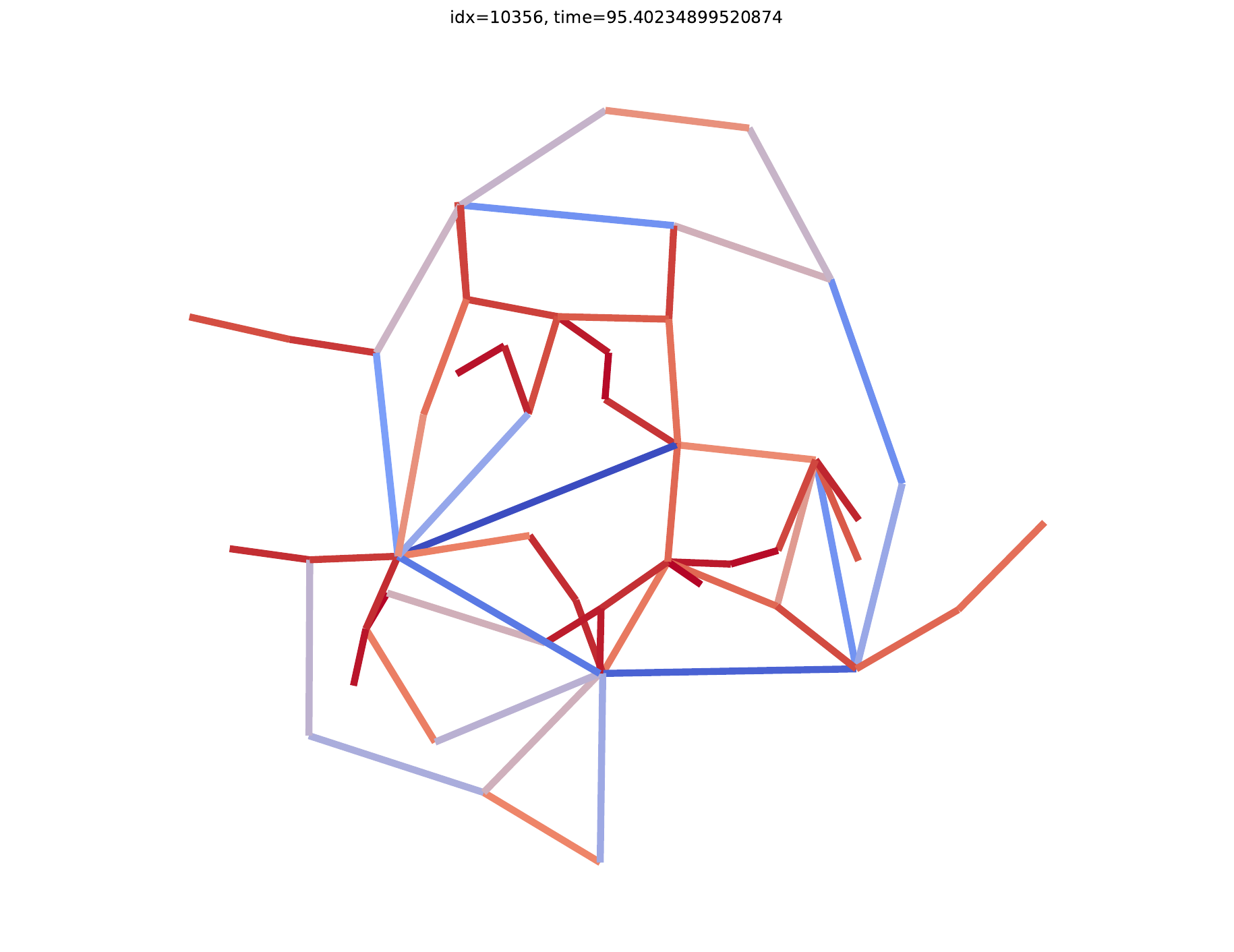} &
\imgcell{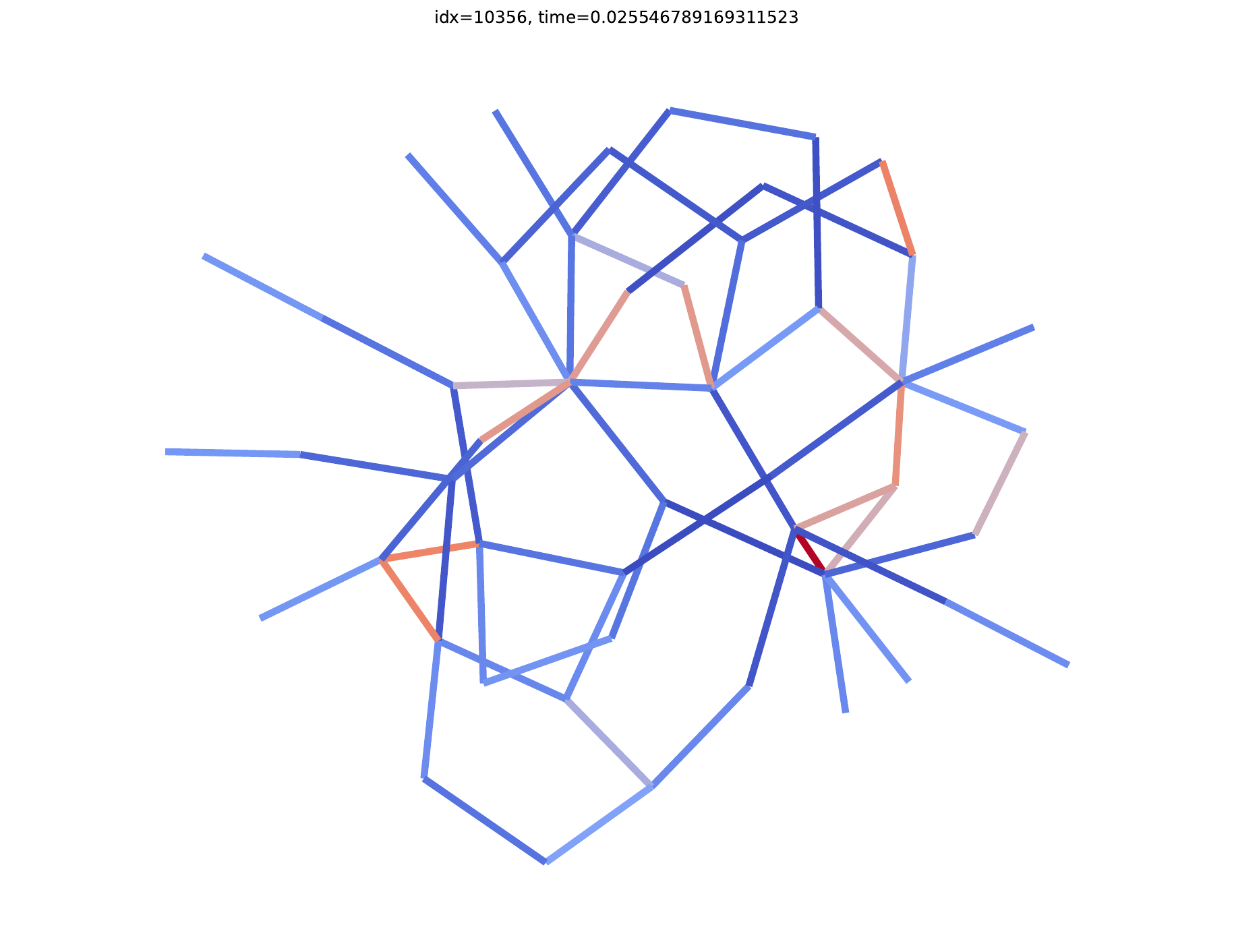} &
\imgcell{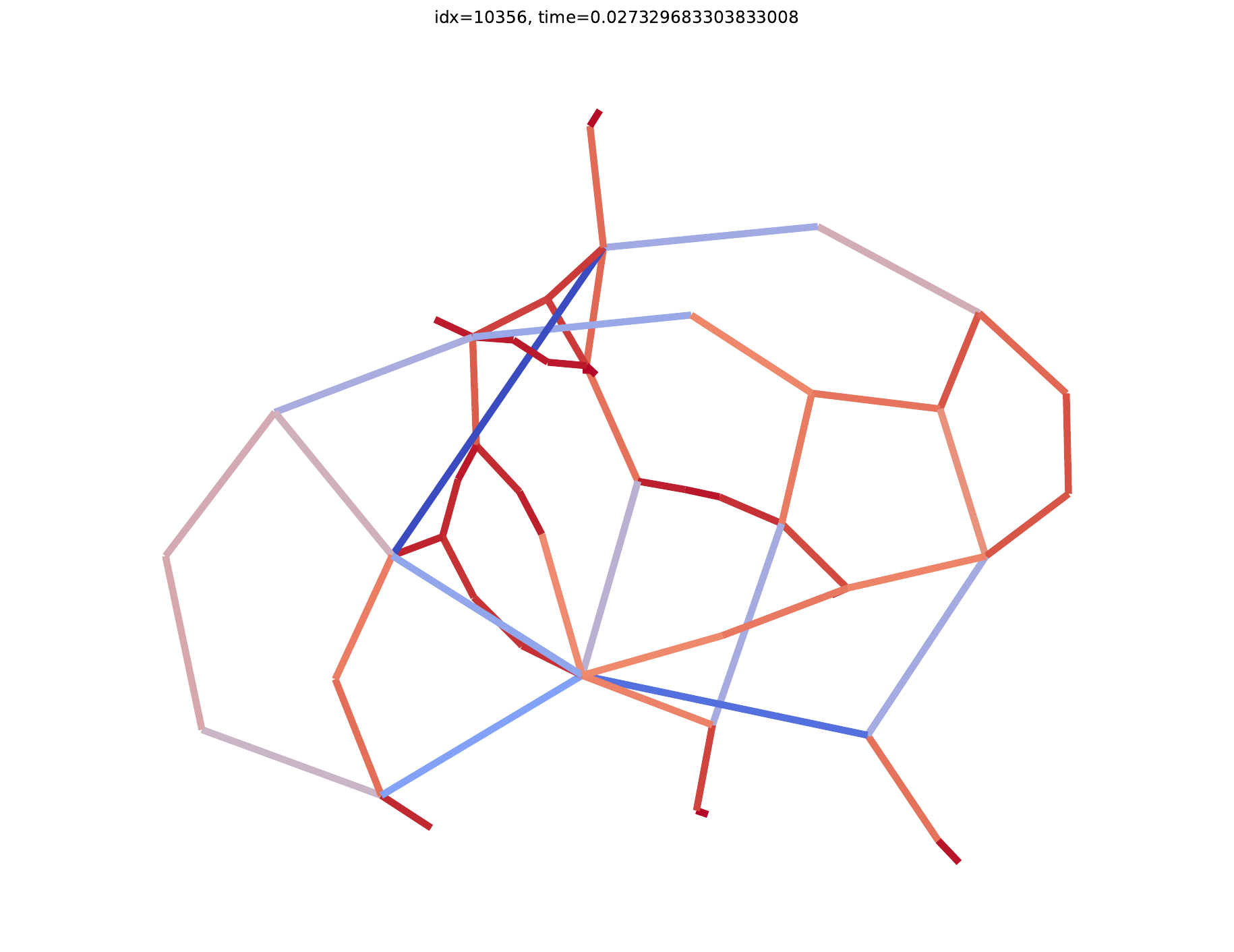} &
\imgcell{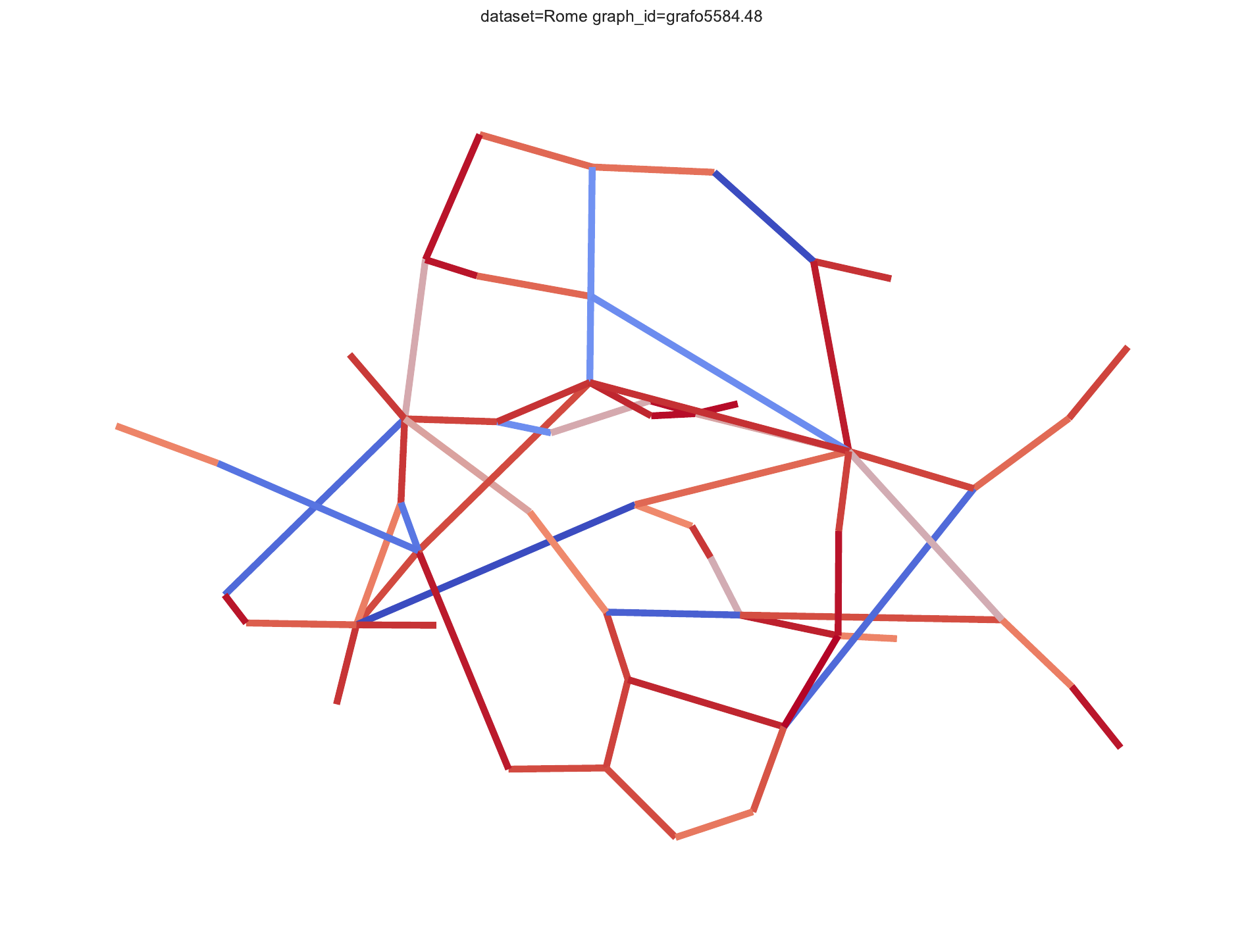} &
\imgcell{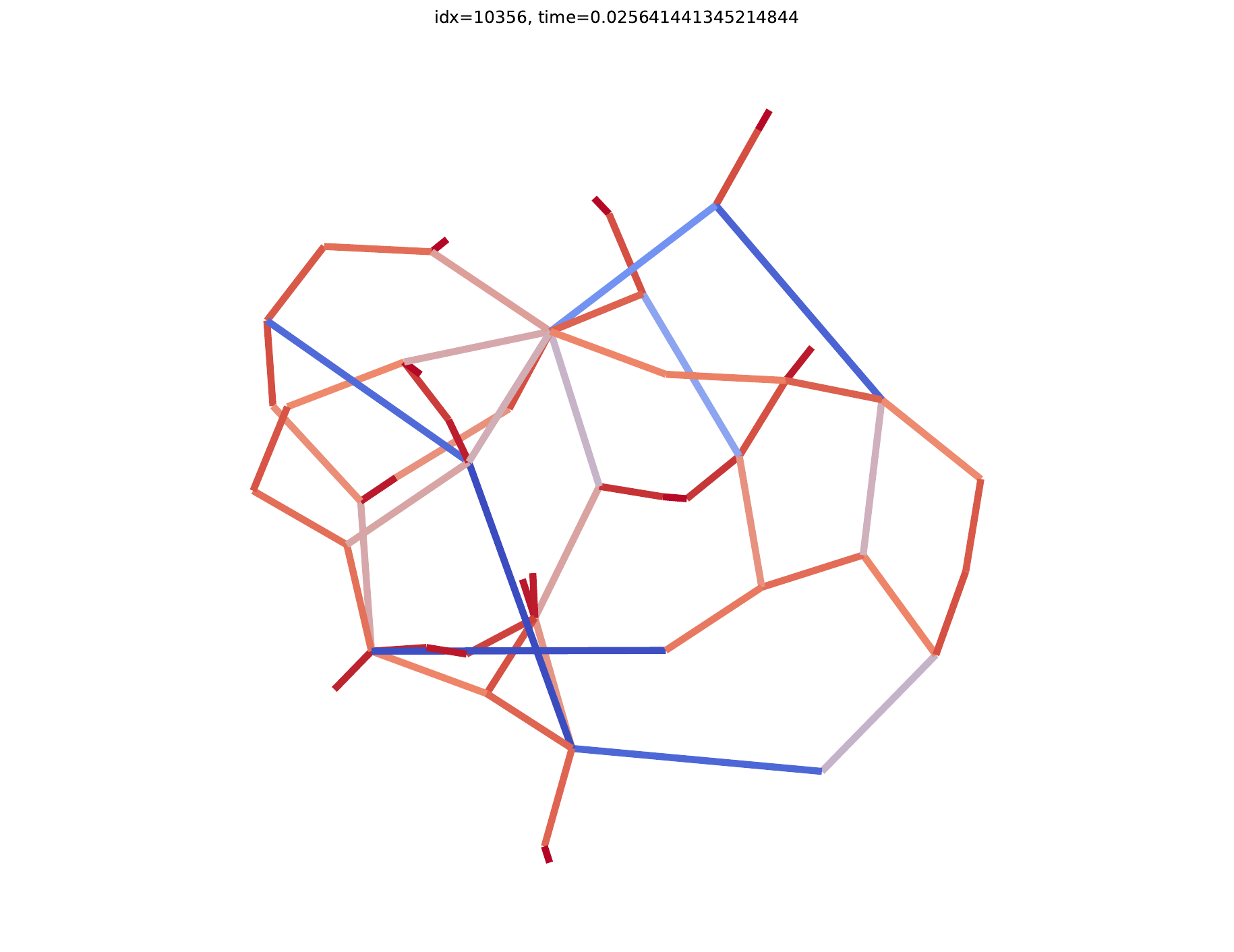} &
\imgcell{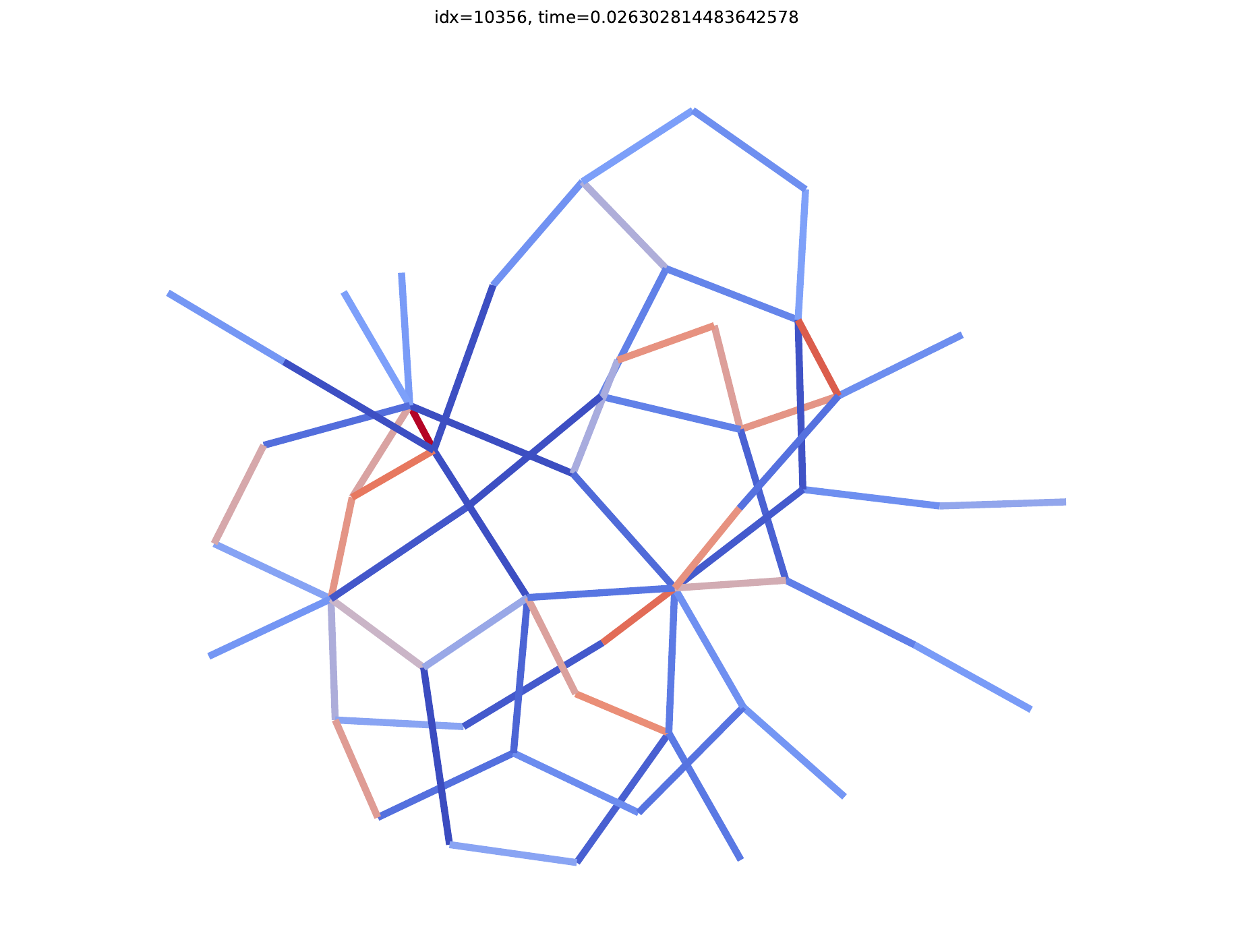} &
\imgcell{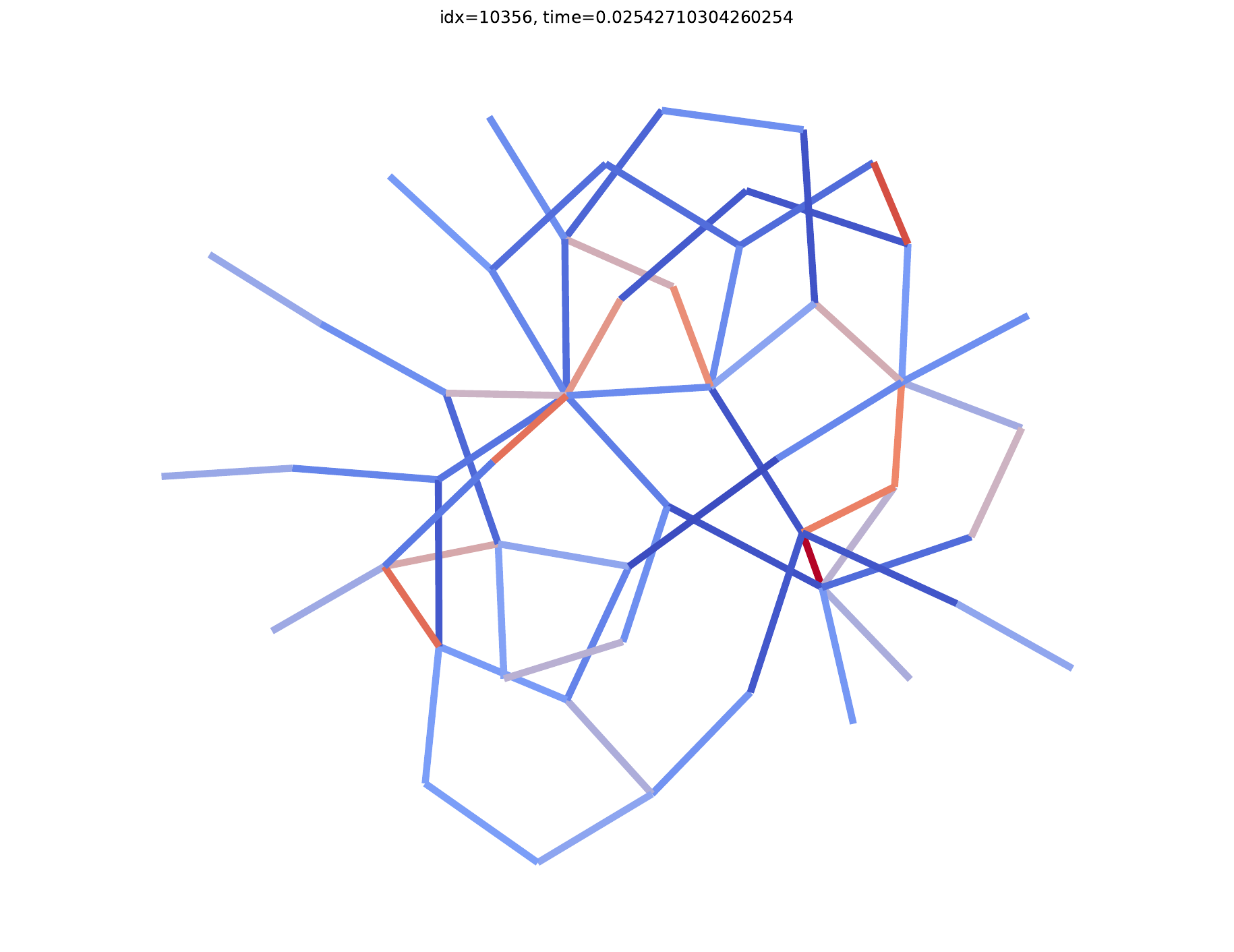} &
\imgcell{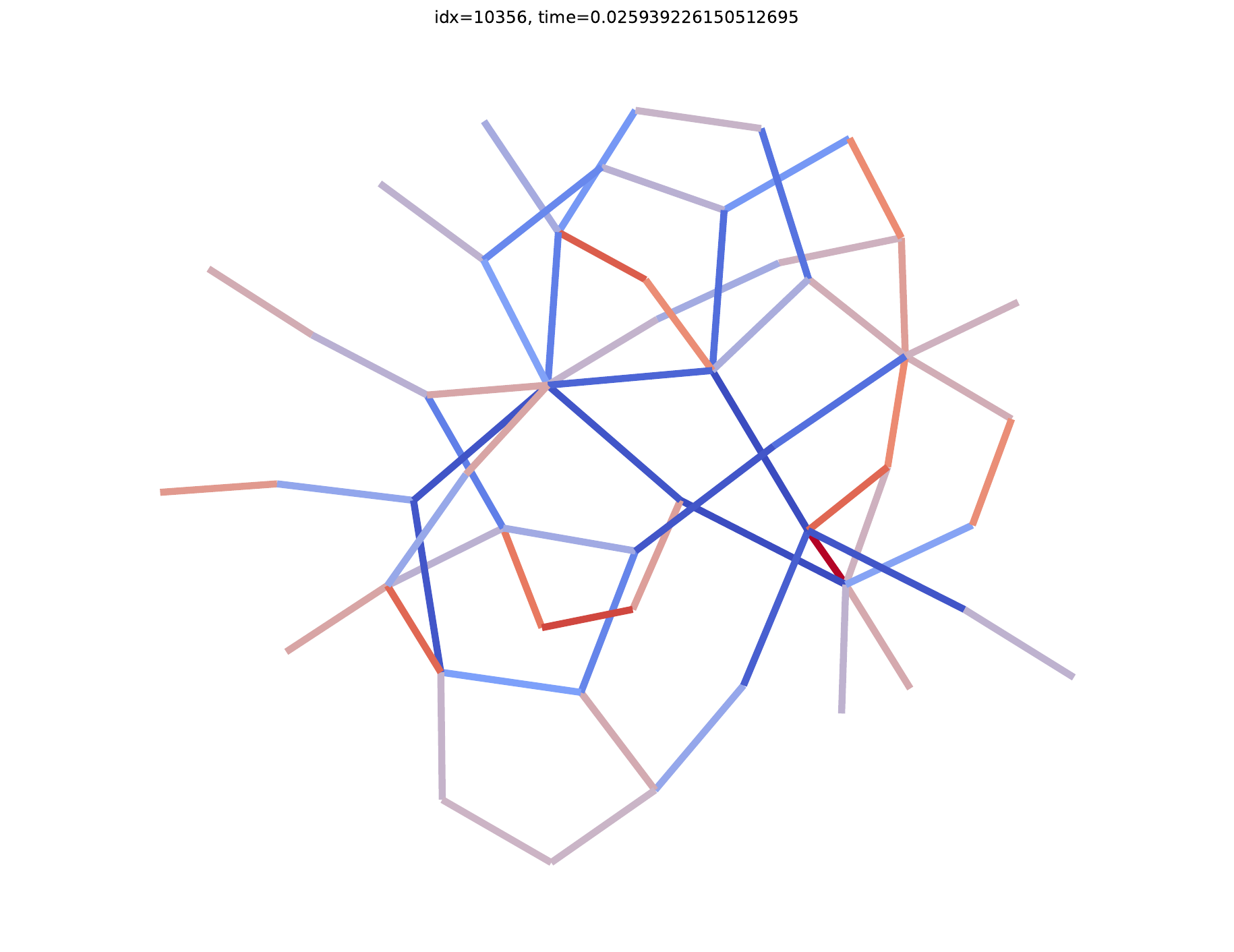} \\

&
t = 0.00s &
t = 0.43s &
t = 0.18s &
t = 0.05s &
t = 95.40s &
t = 0.03s &
t = 0.03s &
t = 0.03s &
t = 0.03s &
t = 0.03s &
t = 0.03s &
t = 0.03s \\

\makecell{\bfseries grafo1925.35\\N = 38\\M = 51} &
\imgcell{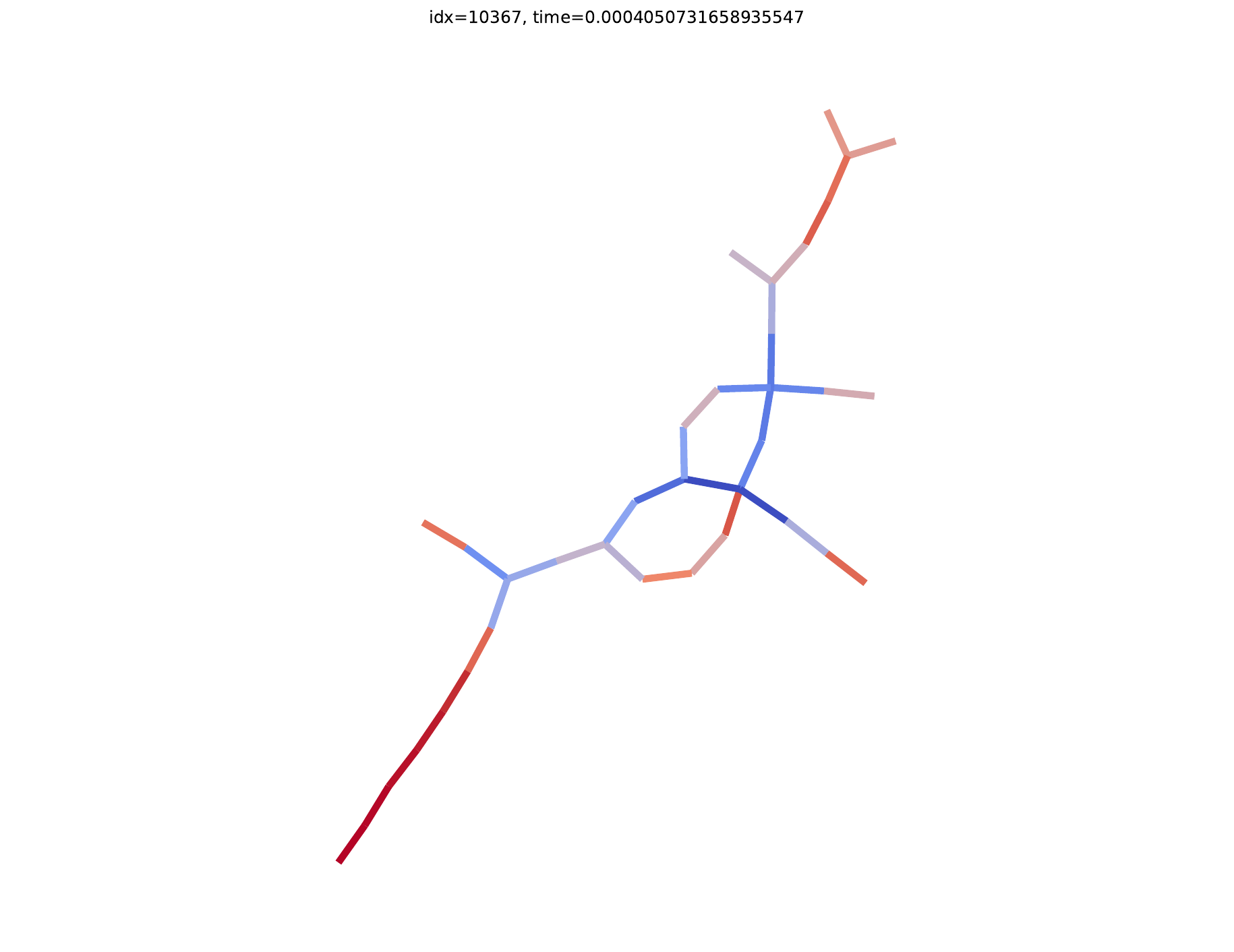} &
\imgcell{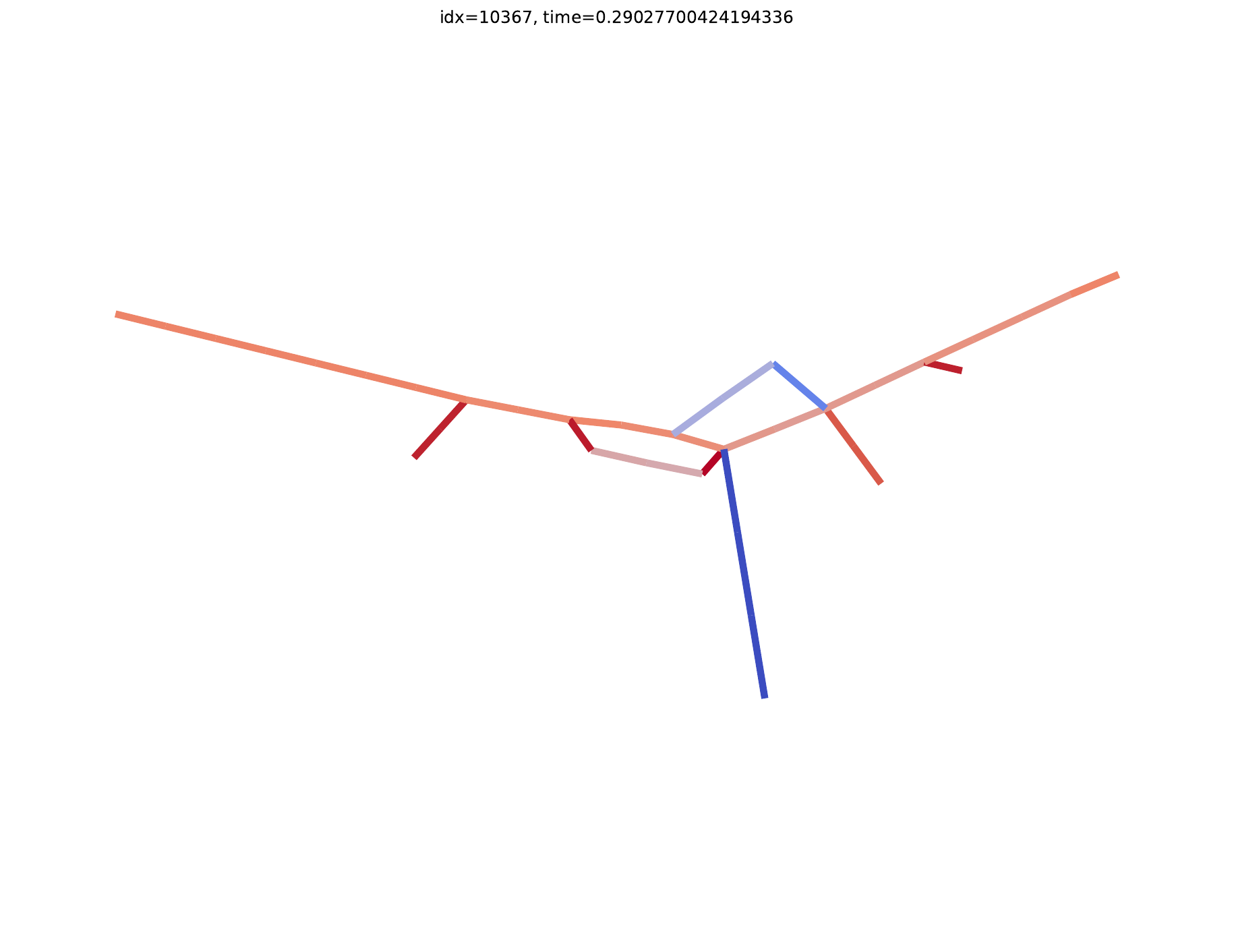} &
\imgcell{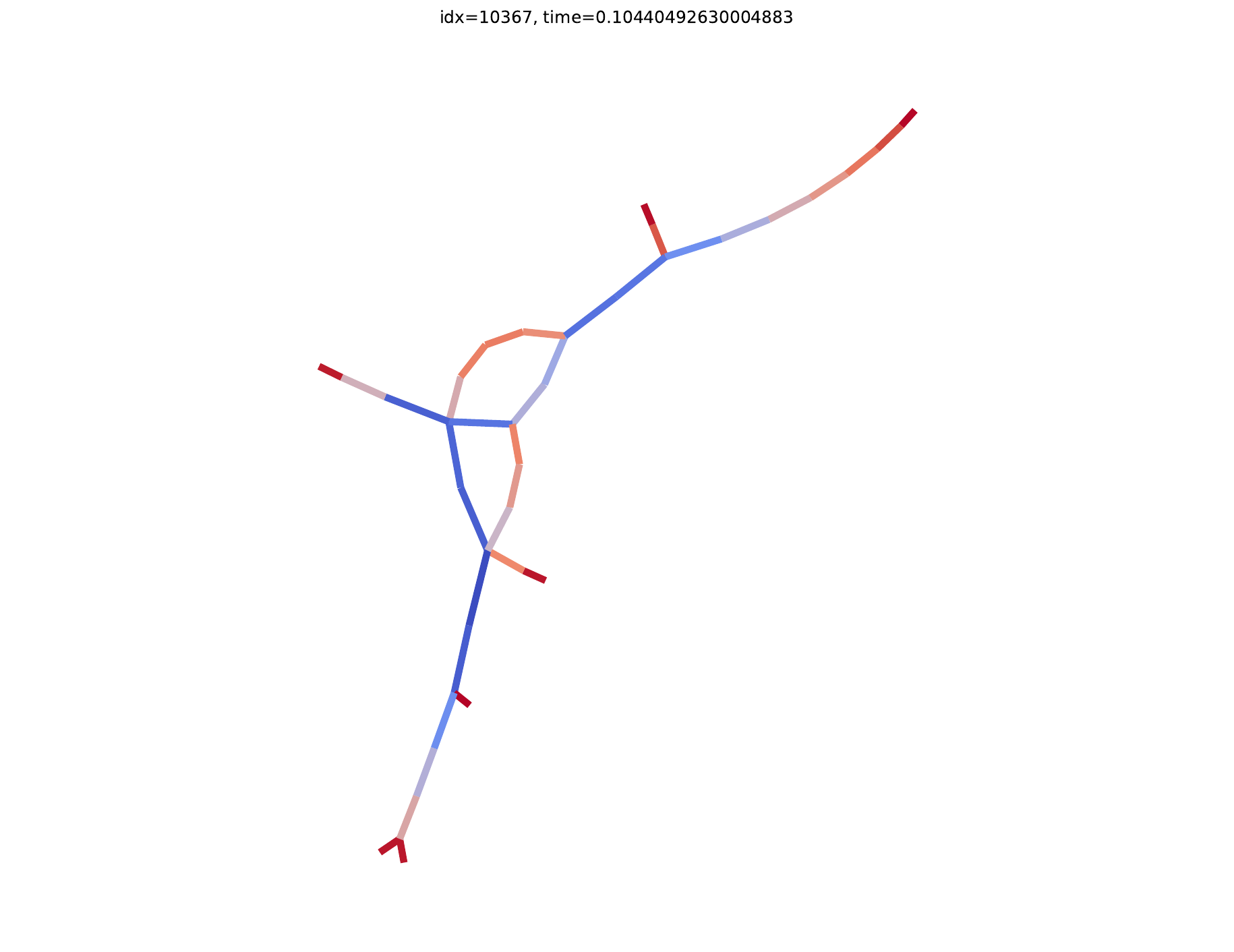} &
\imgcell{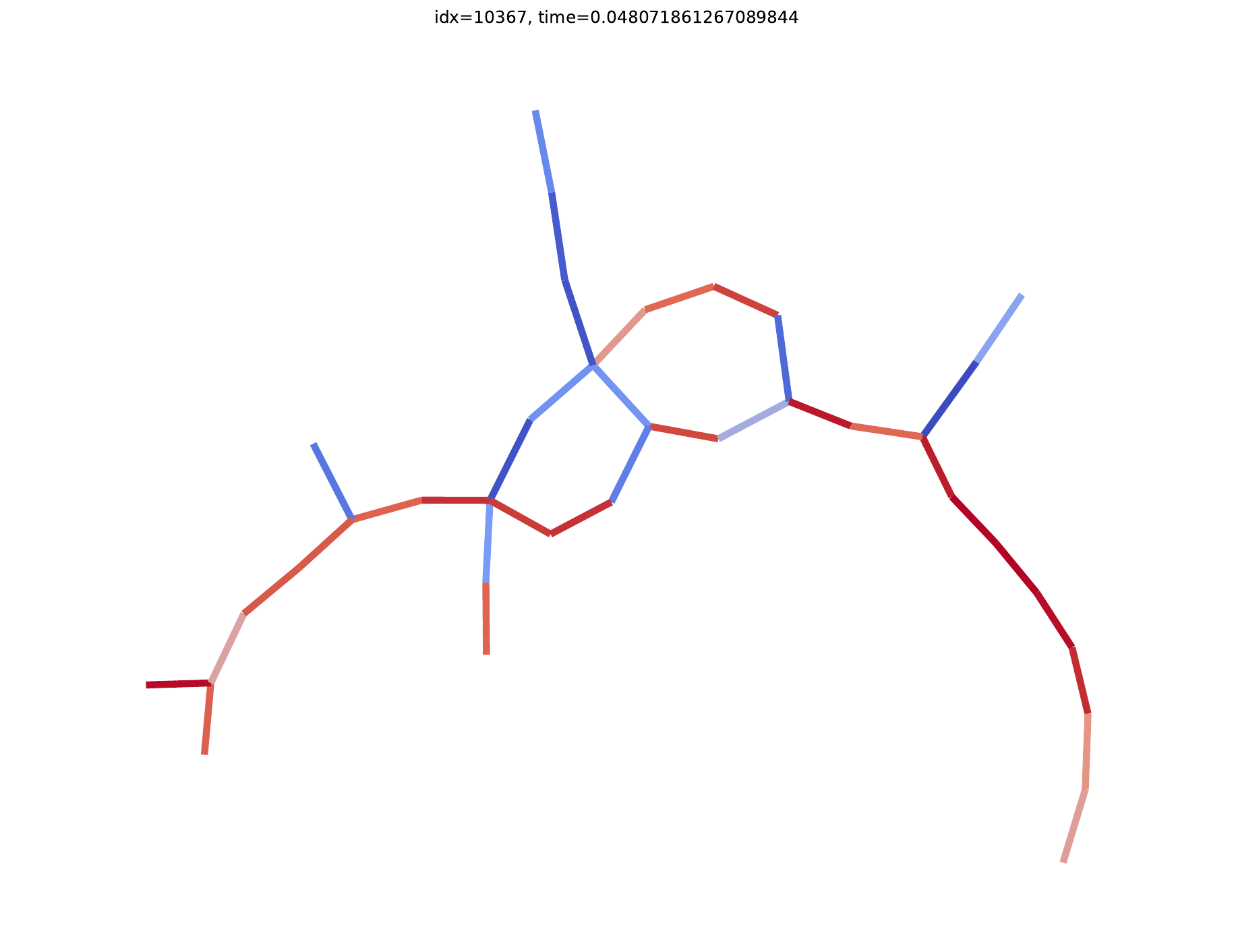} &
\imgcell{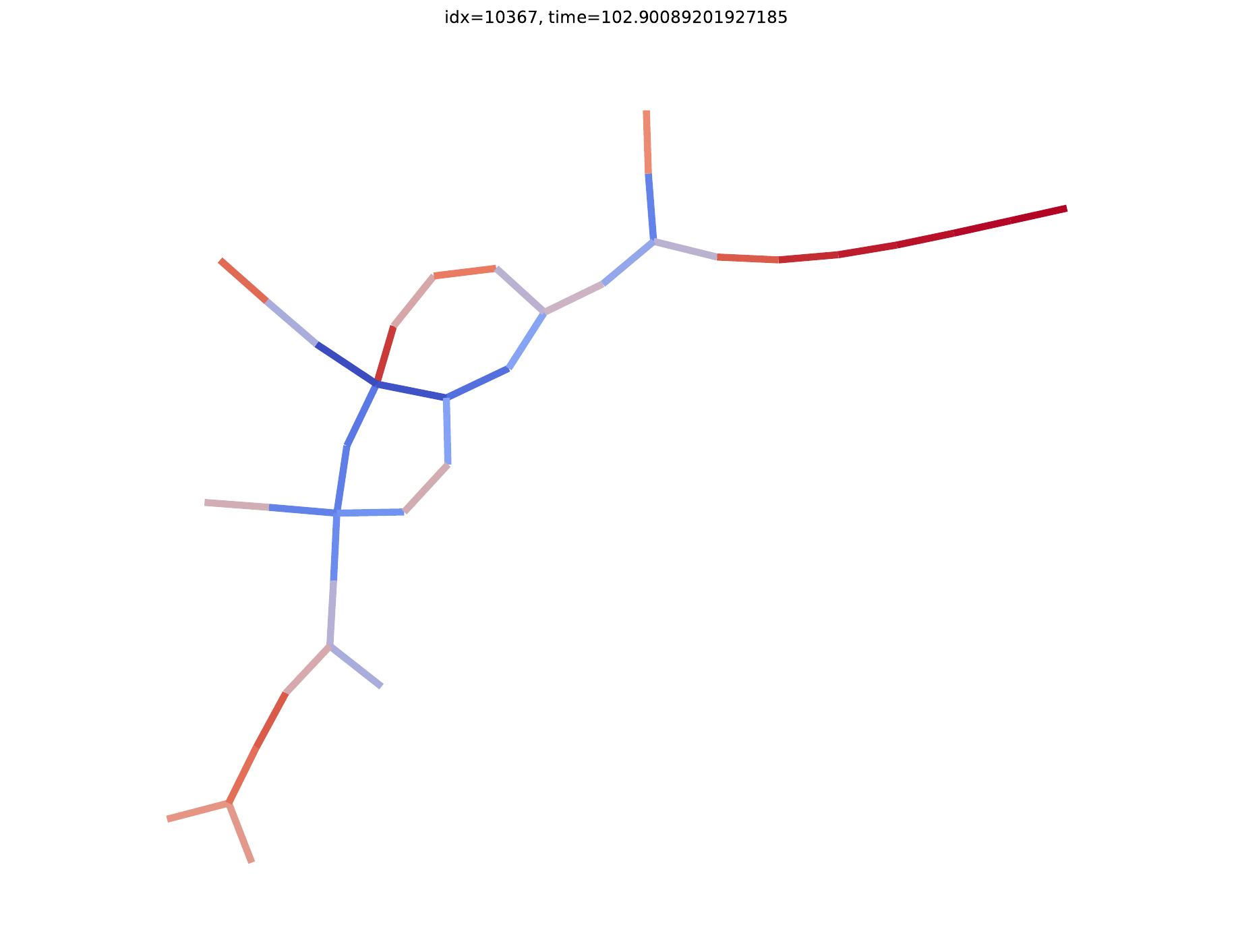} &
\imgcell{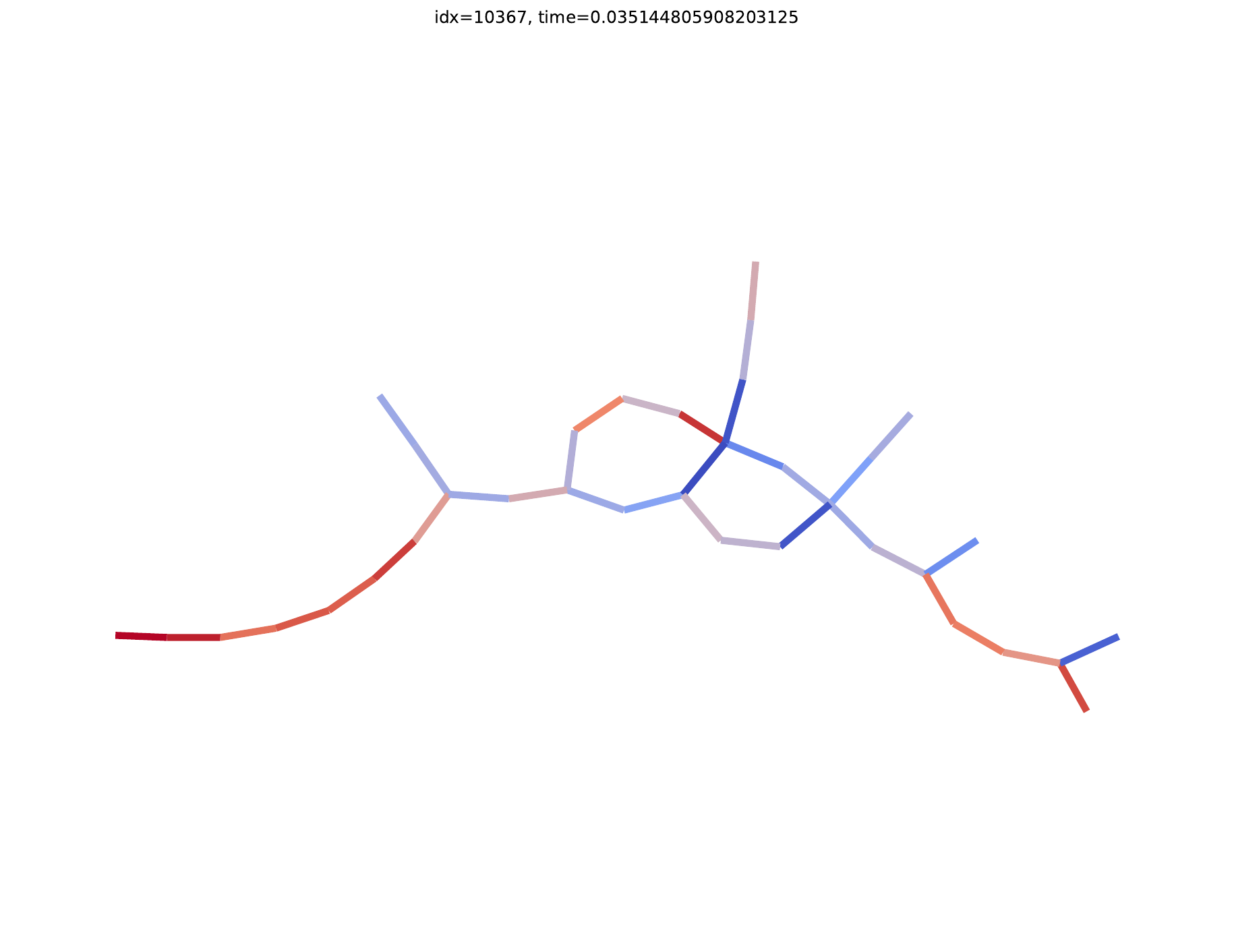} &
\imgcell{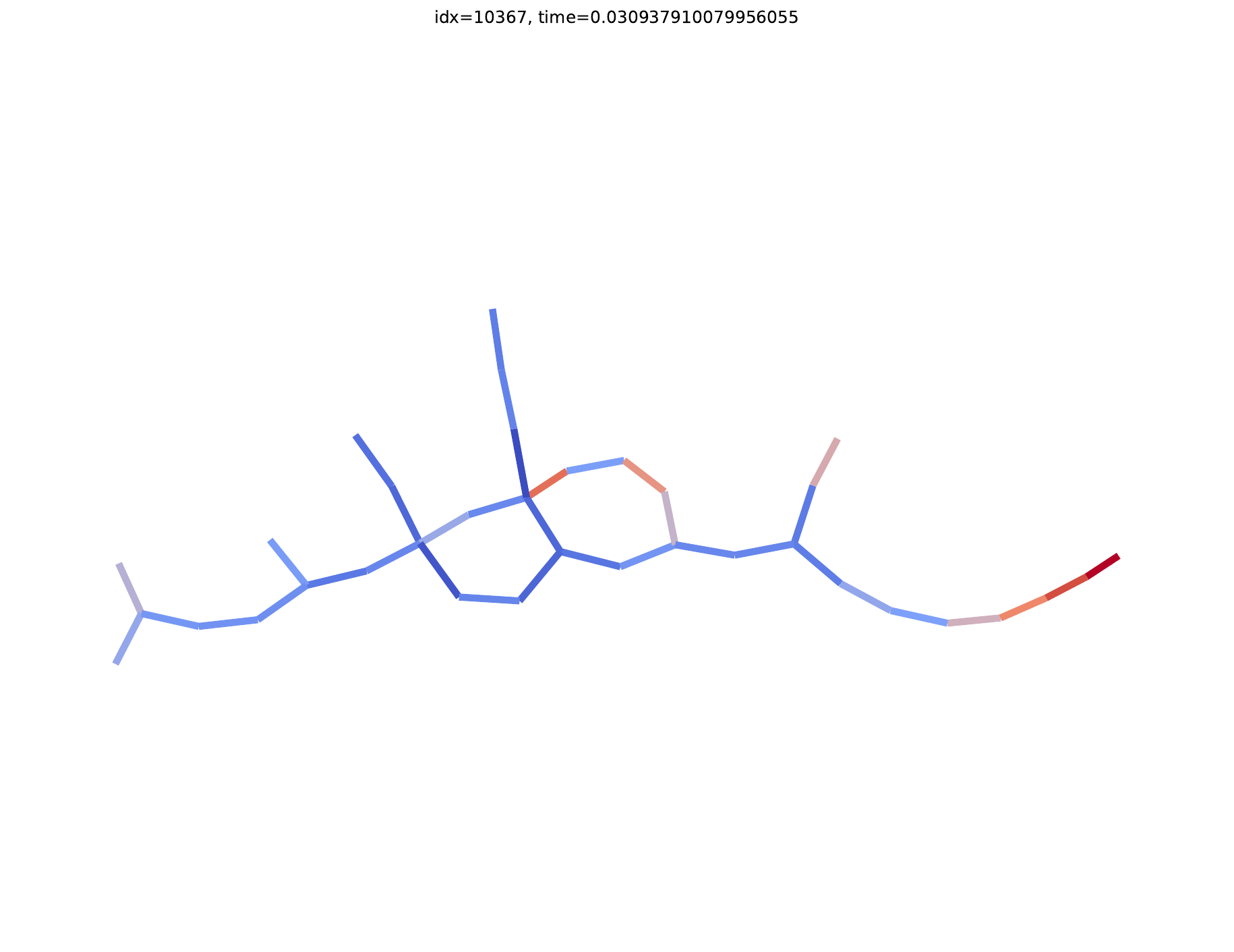} &
\imgcell{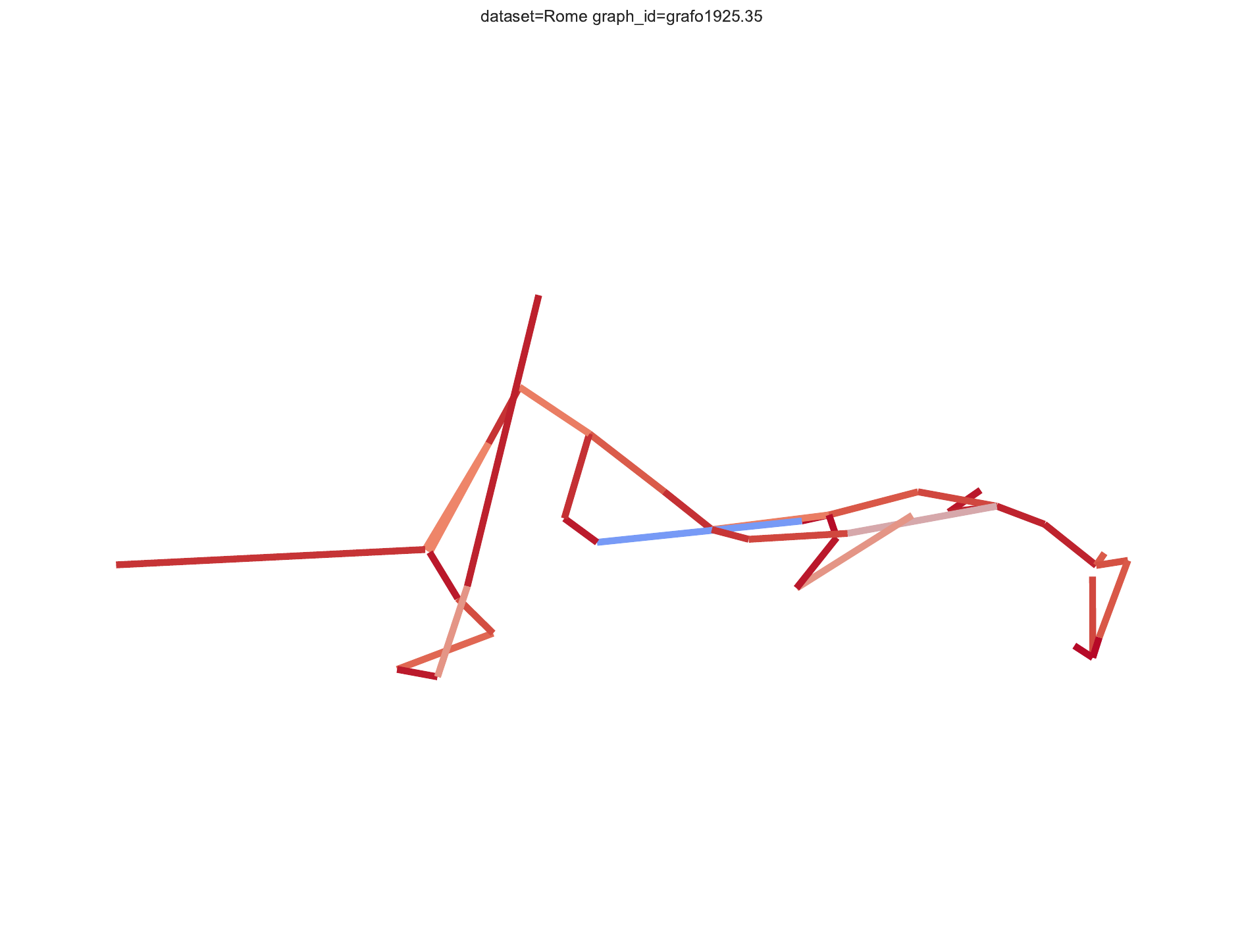} &
\imgcell{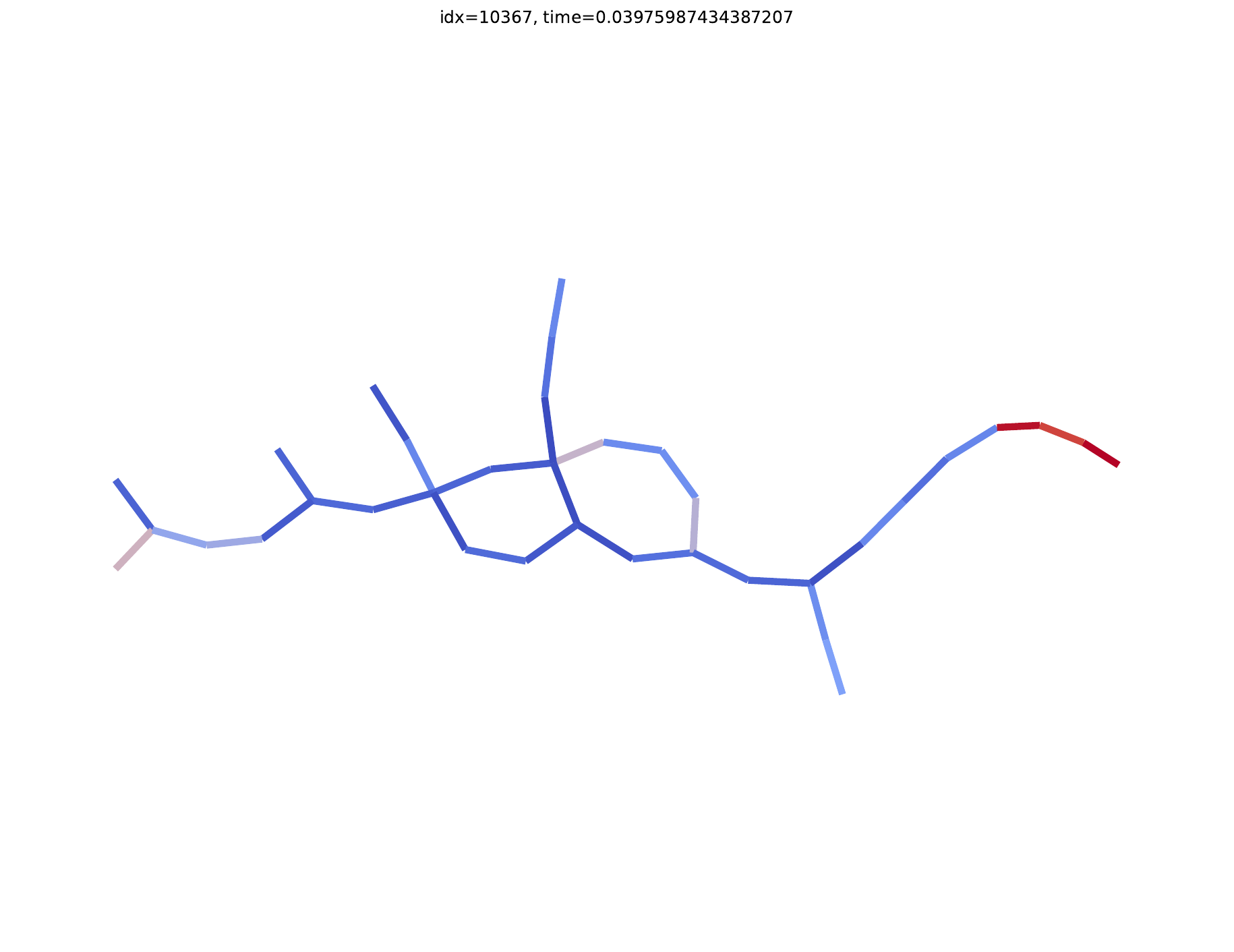} &
\imgcell{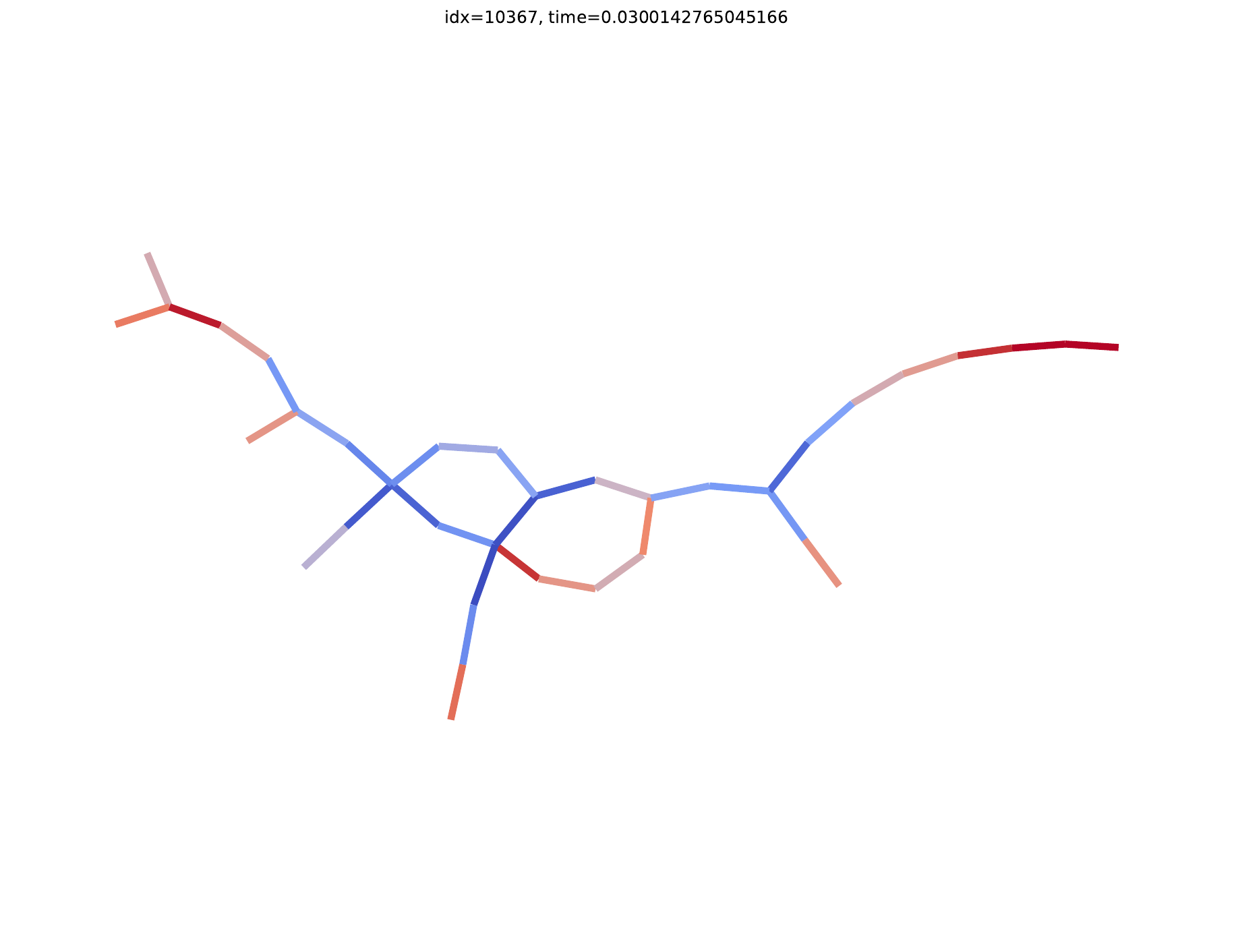} &
\imgcell{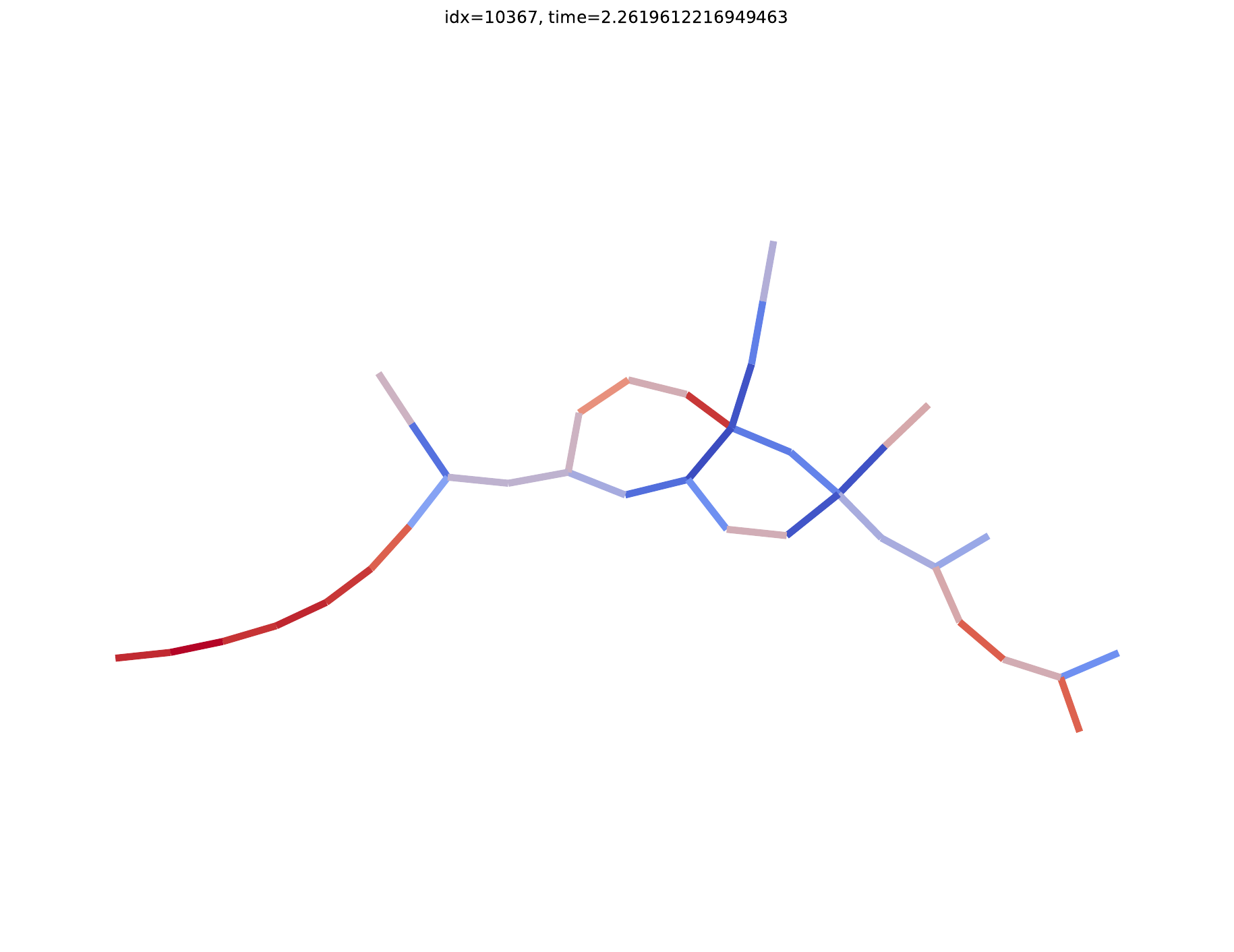} &
\imgcell{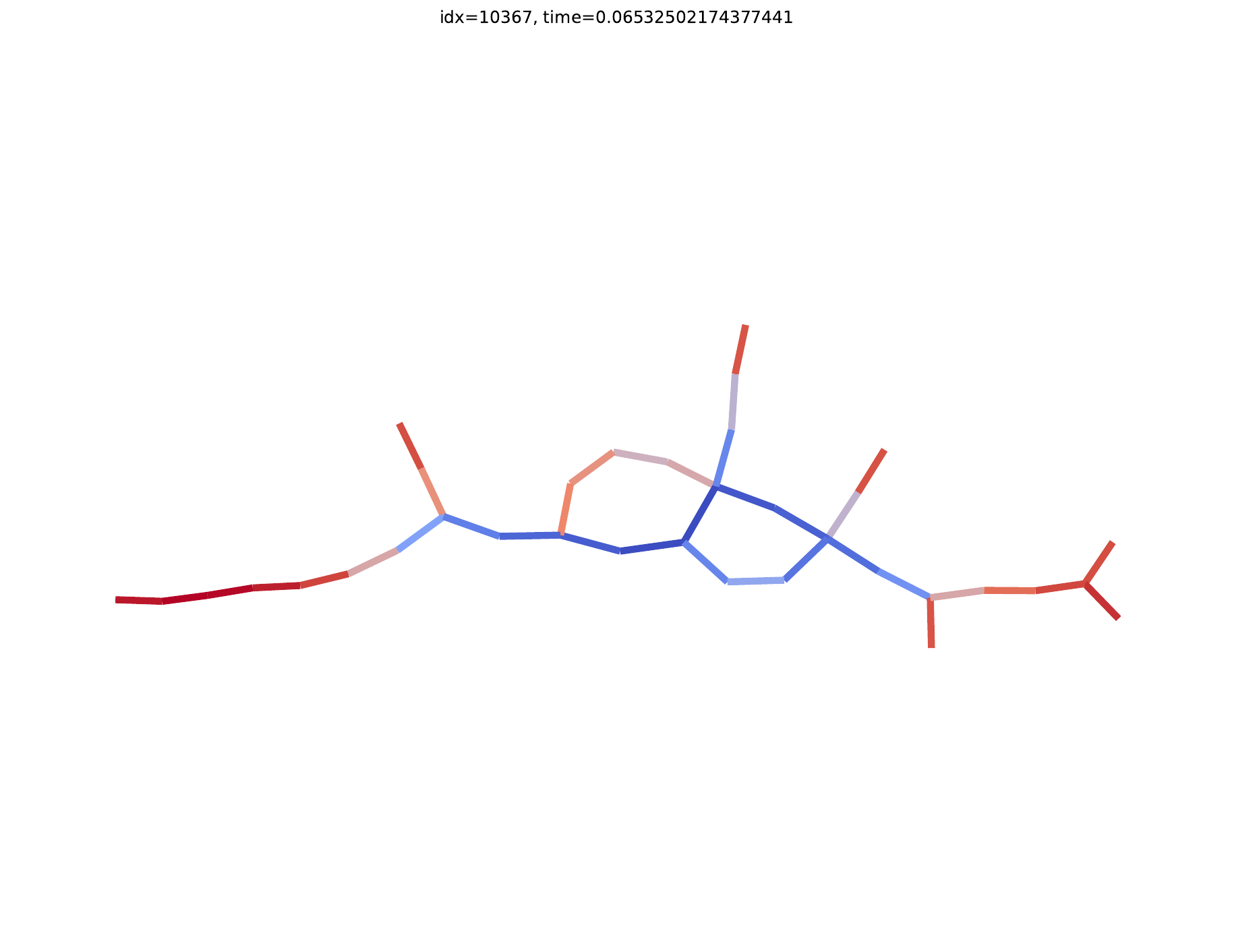} \\

&
t = 0.00s &
t = 0.29s &
t = 0.10s &
t = 0.05s &
t = 102.90s &
t = 0.04s &
t = 0.03s &
t = 0.04s &
t = 0.04s &
t = 0.03s &
t = 0.06s &
t = 0.07s \\

\makecell{\bfseries grafo419.15\\N = 23\\M = 26} &
\imgcell{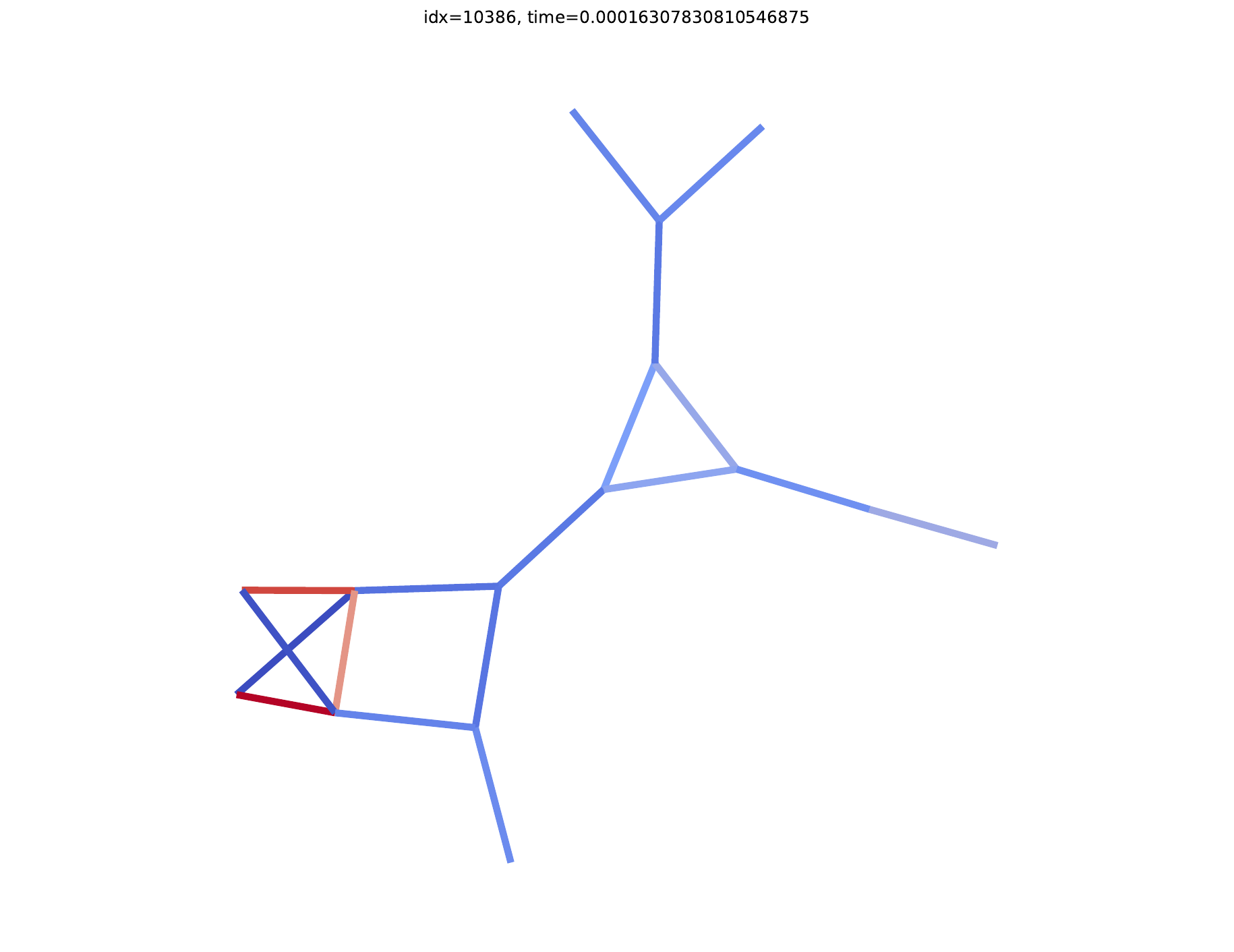} &
\imgcell{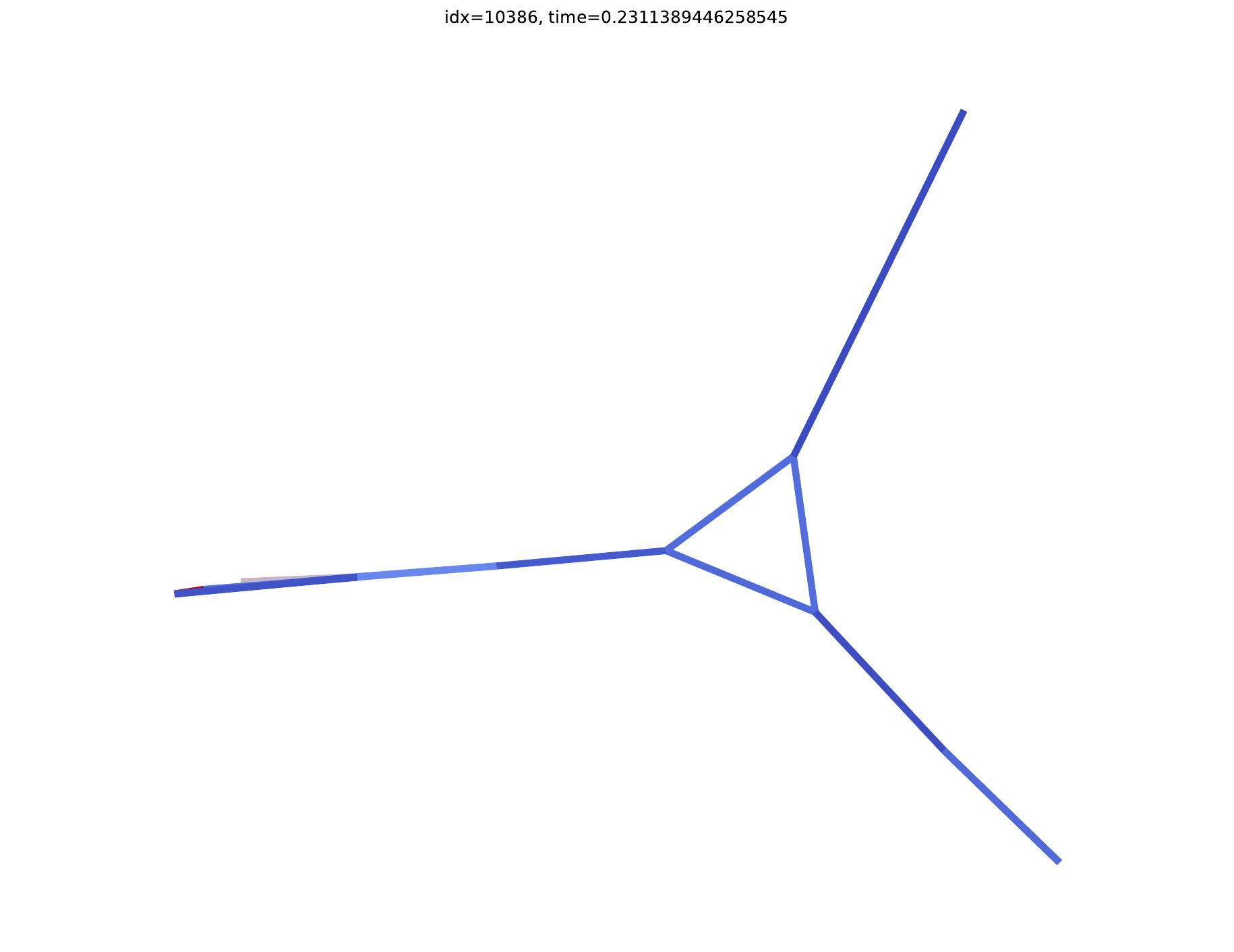} &
\imgcell{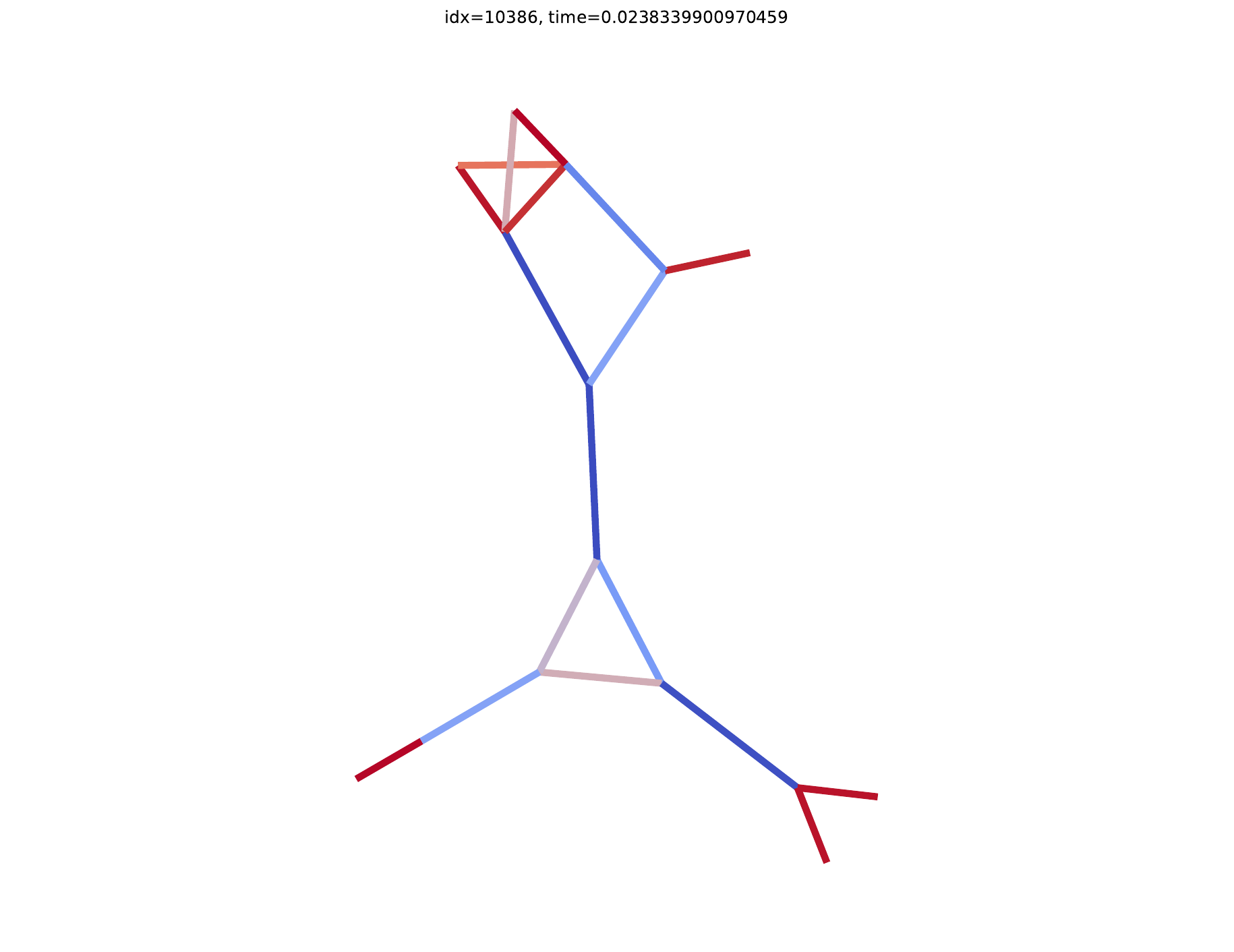} &
\imgcell{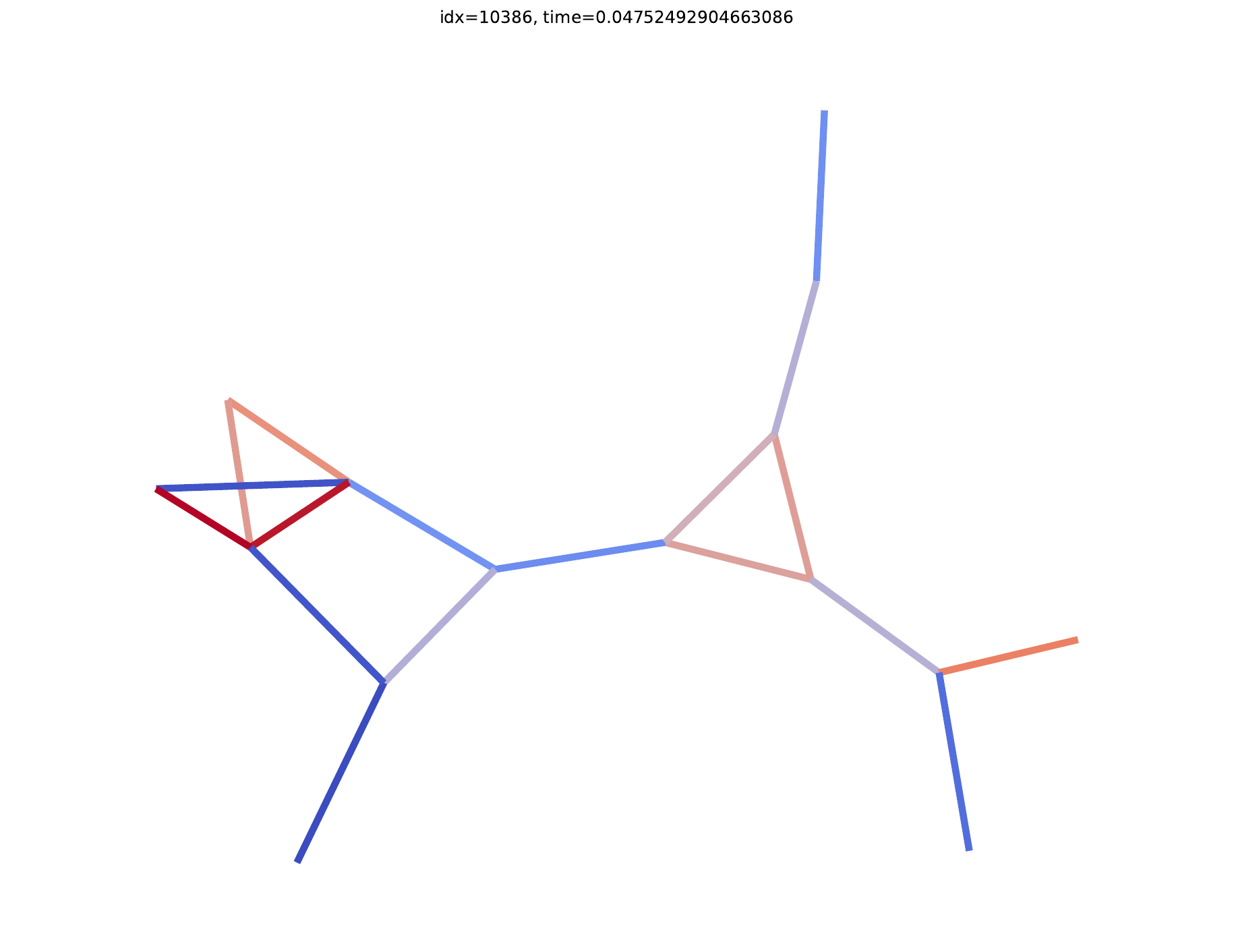} &
\imgcell{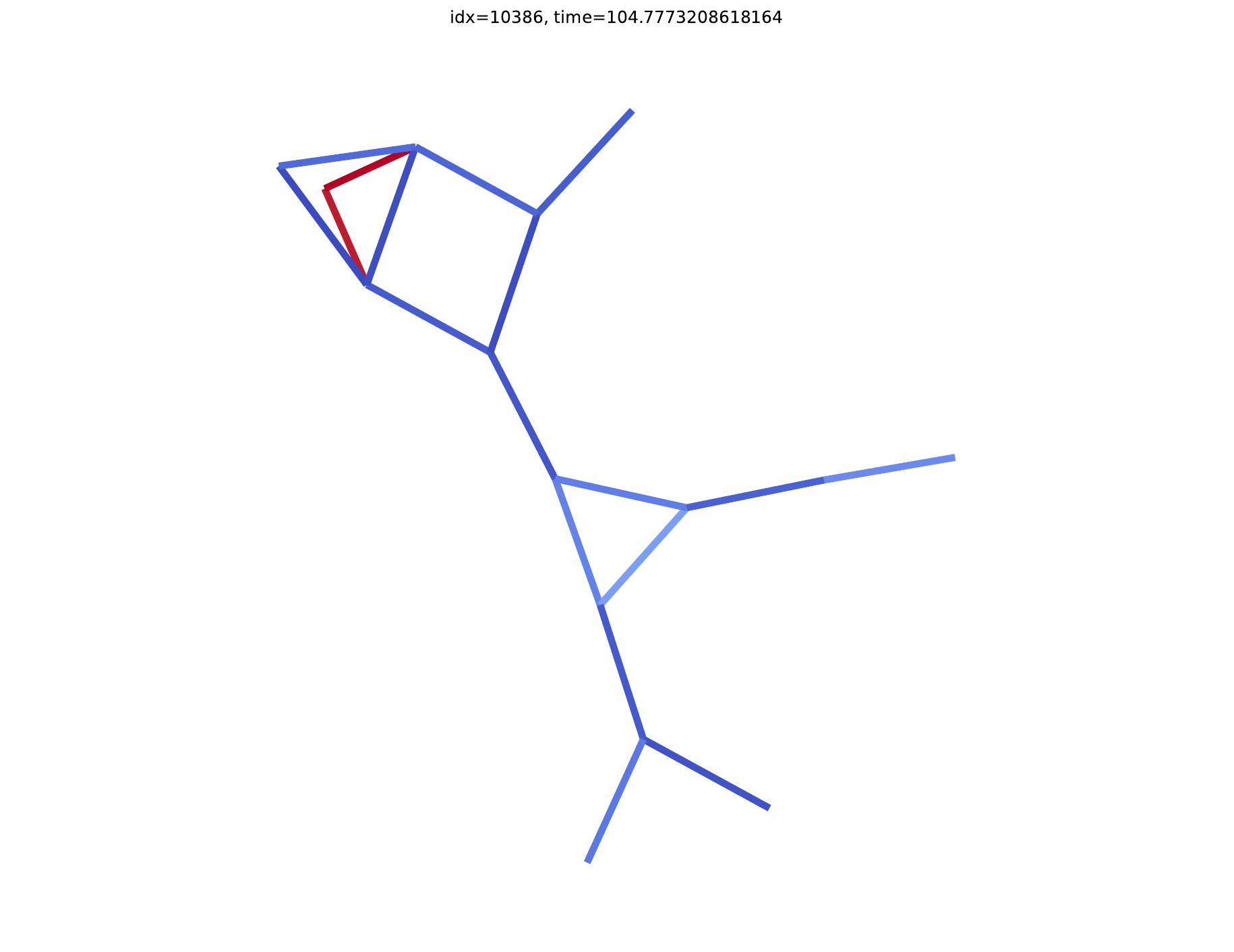} &
\imgcell{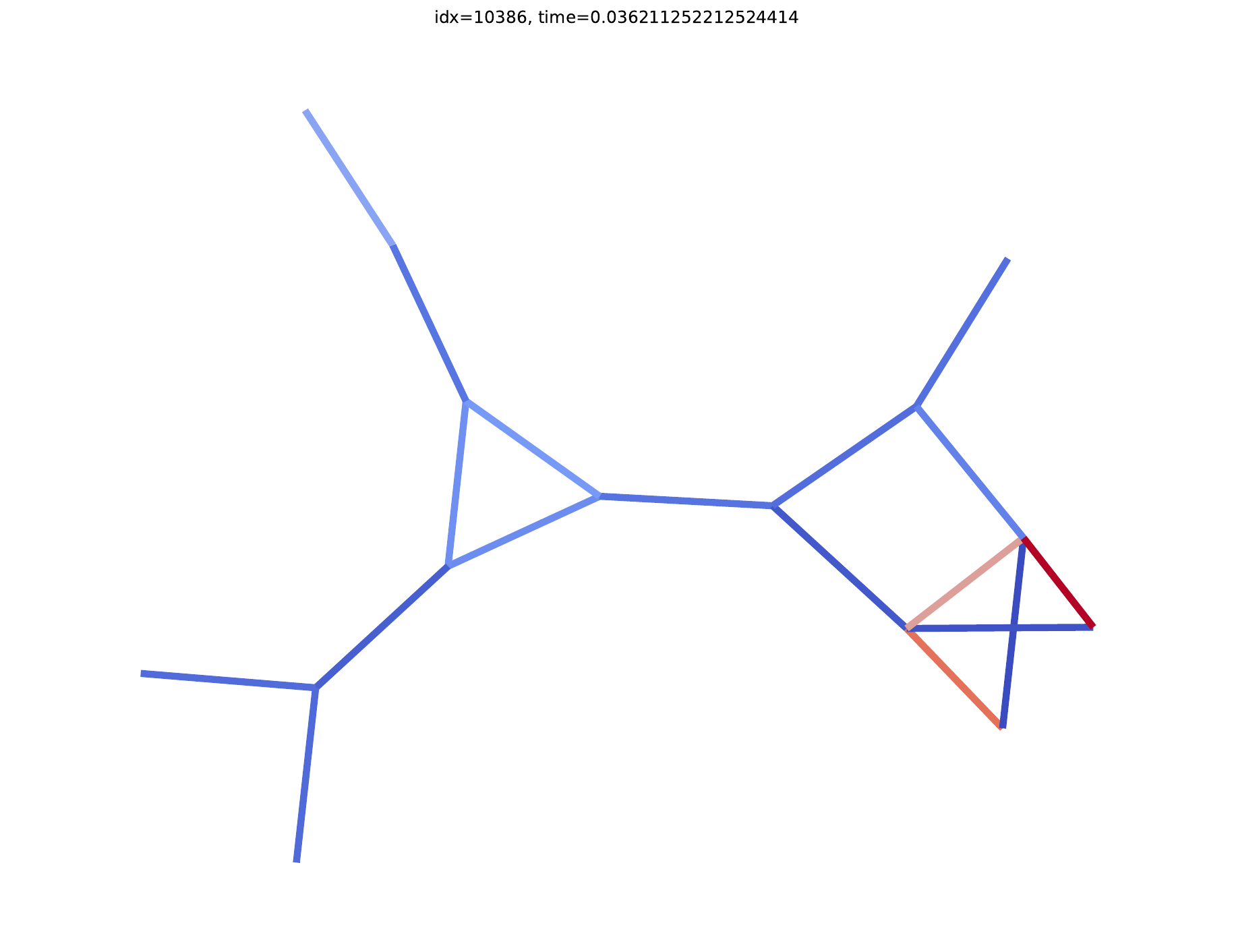} &
\imgcell{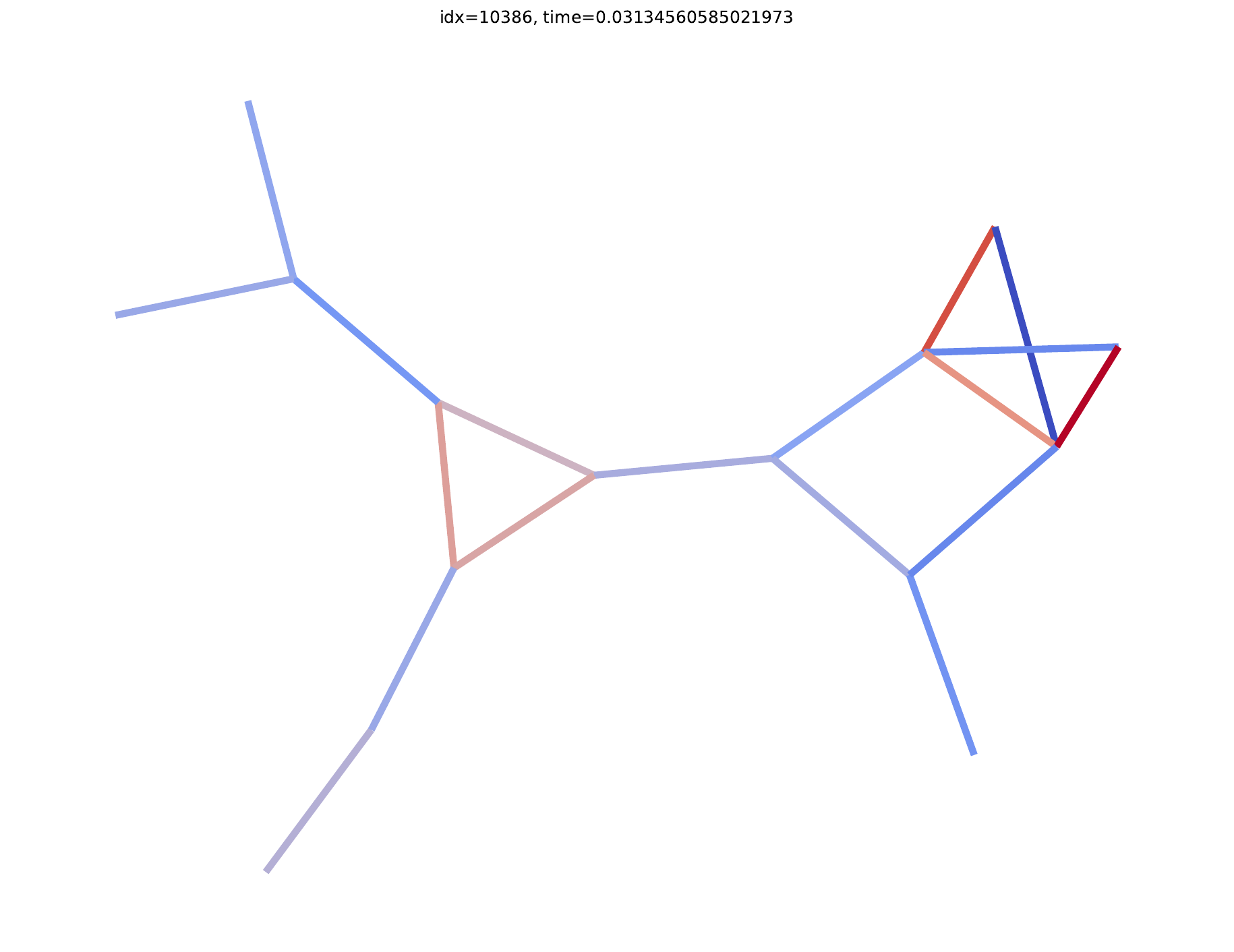} &
\imgcell{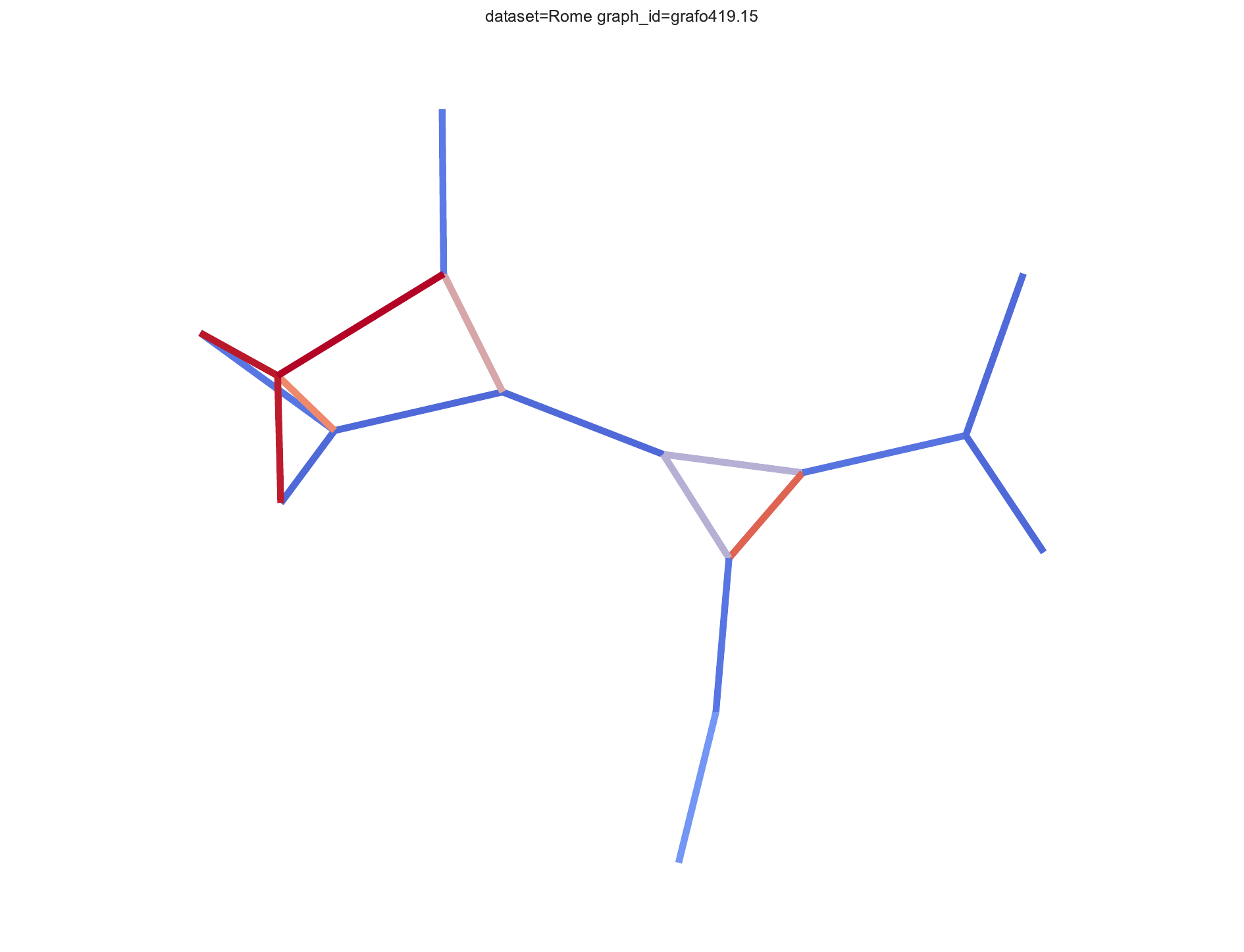} &
\imgcell{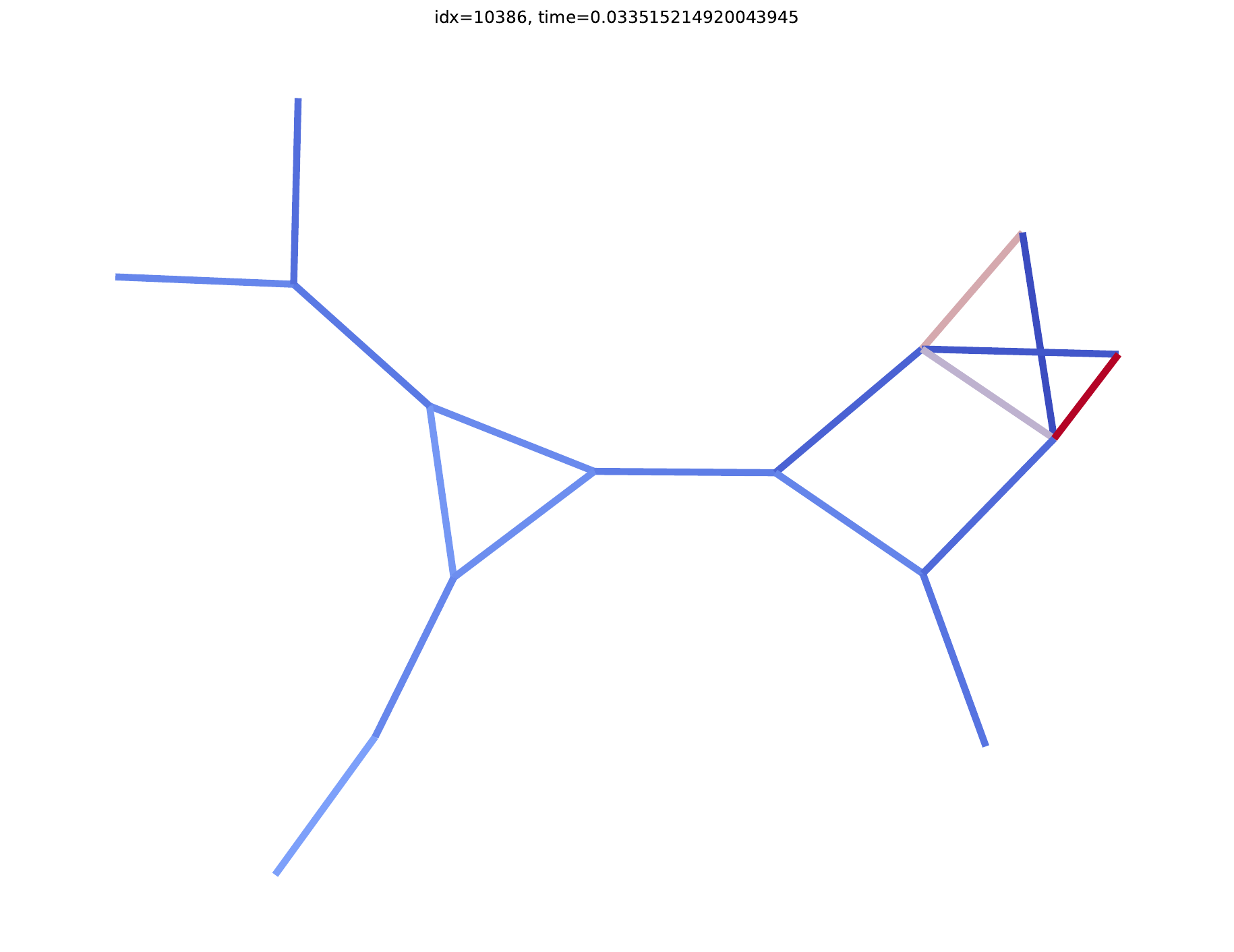} &
\imgcell{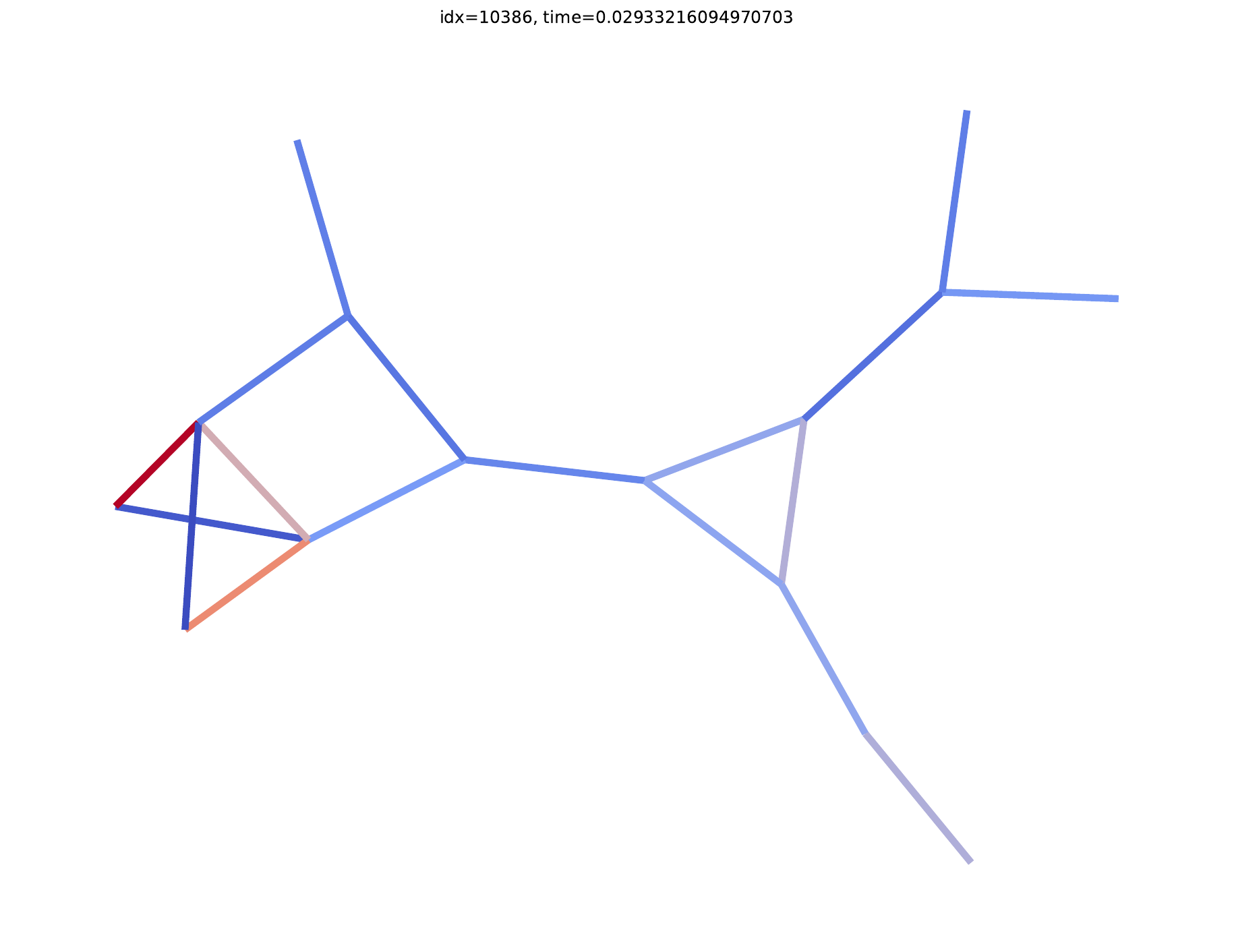} &
\imgcell{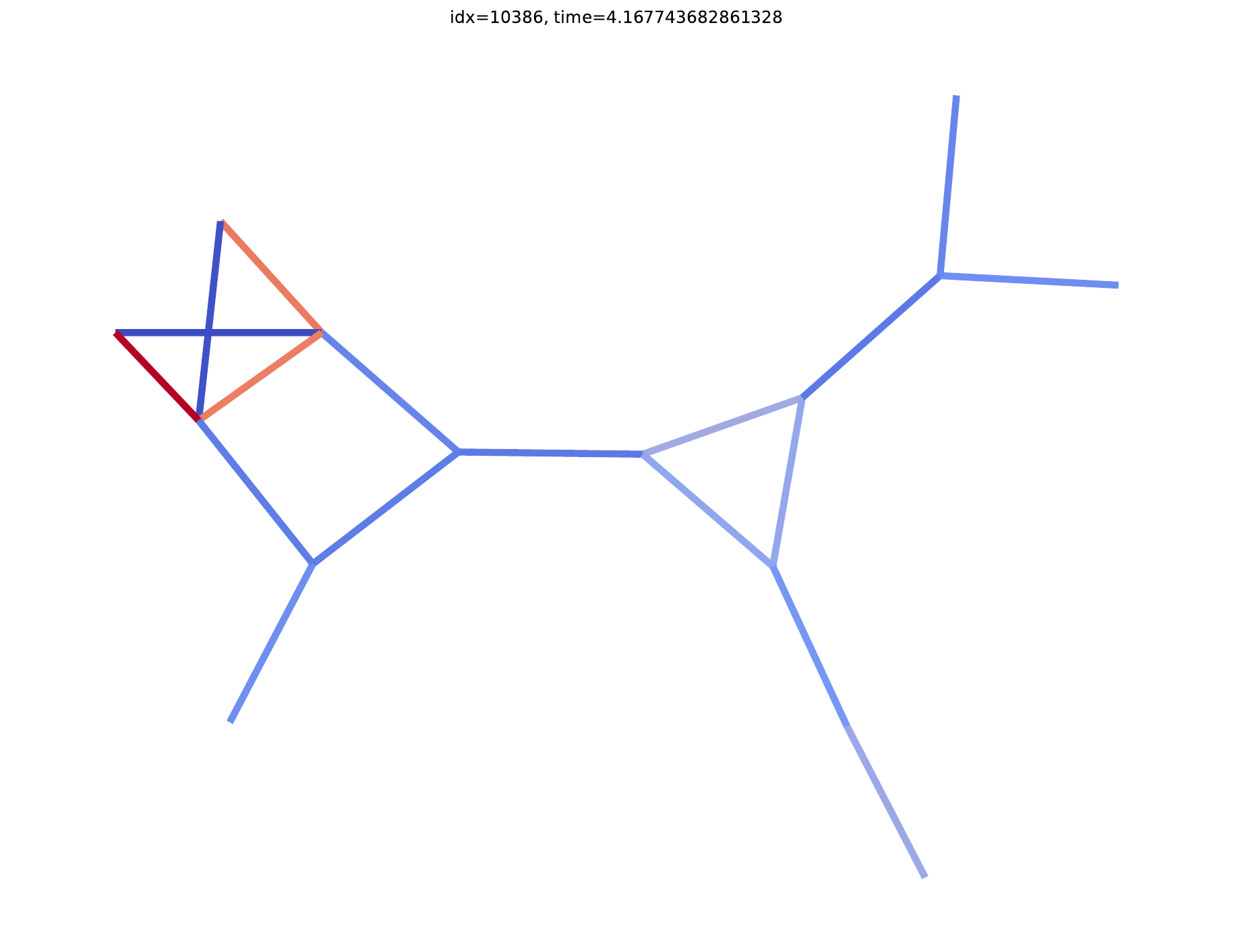} &
\imgcell{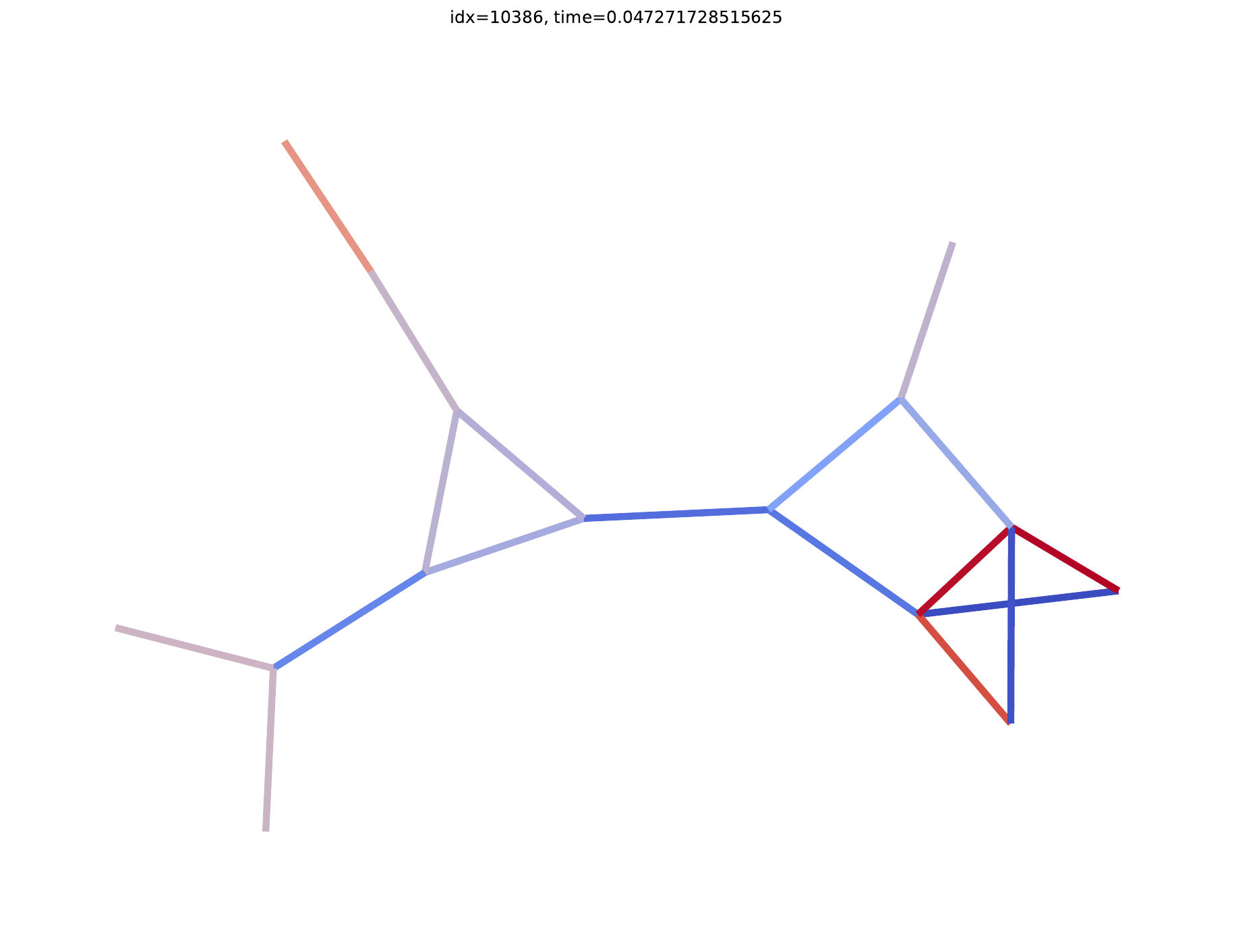} \\

&
t = 0.00s &
t = 0.23s &
t = 0.02s &
t = 0.05s &
t = 104.78s &
t = 0.04s &
t = 0.03s &
t = 0.05s &
t = 0.03s &
t = 0.03s &
t = 0.04s &
t = 0.05s \\

\makecell{\bfseries grafo3411.52\\N = 11\\M = 11} &
\imgcell{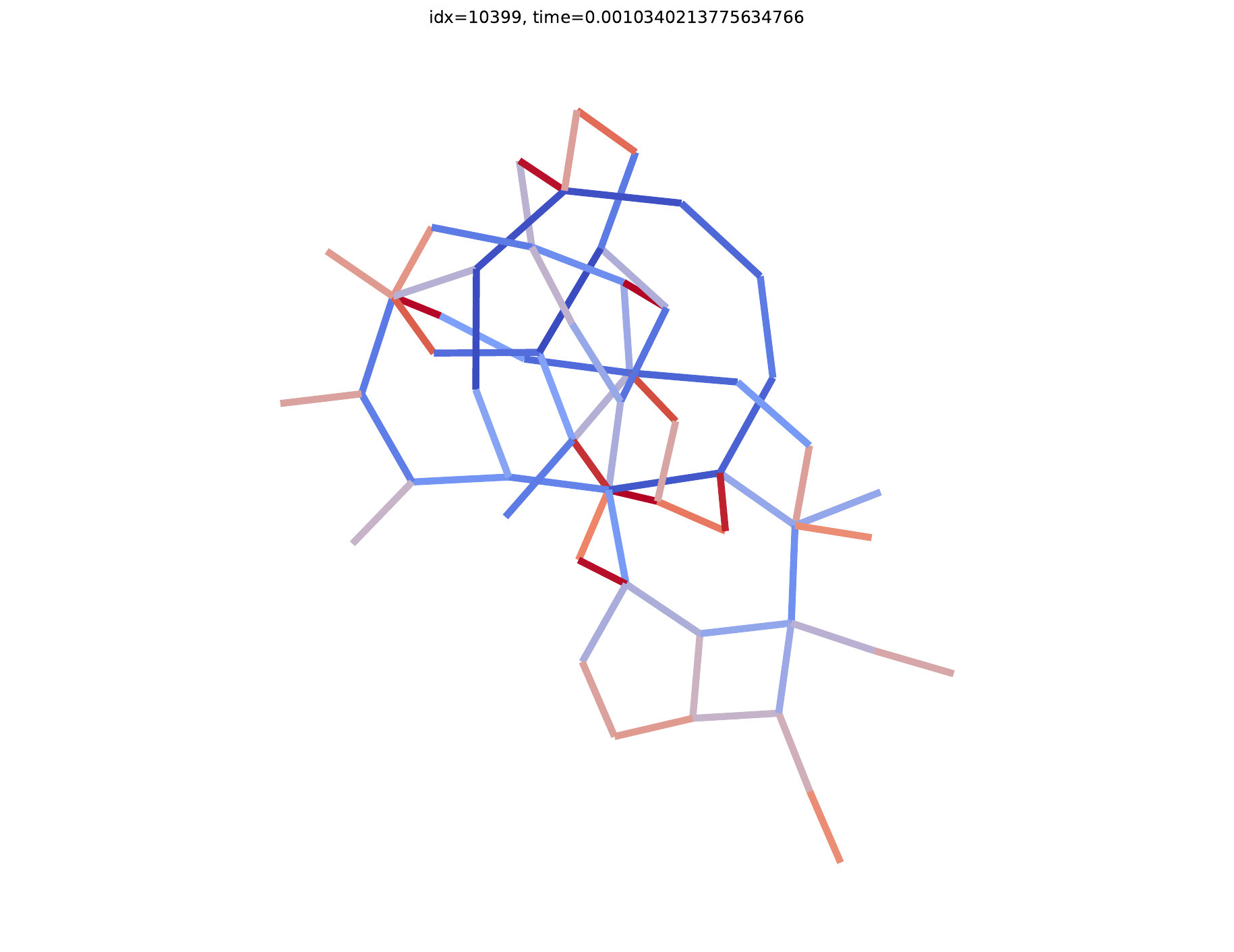} &
\imgcell{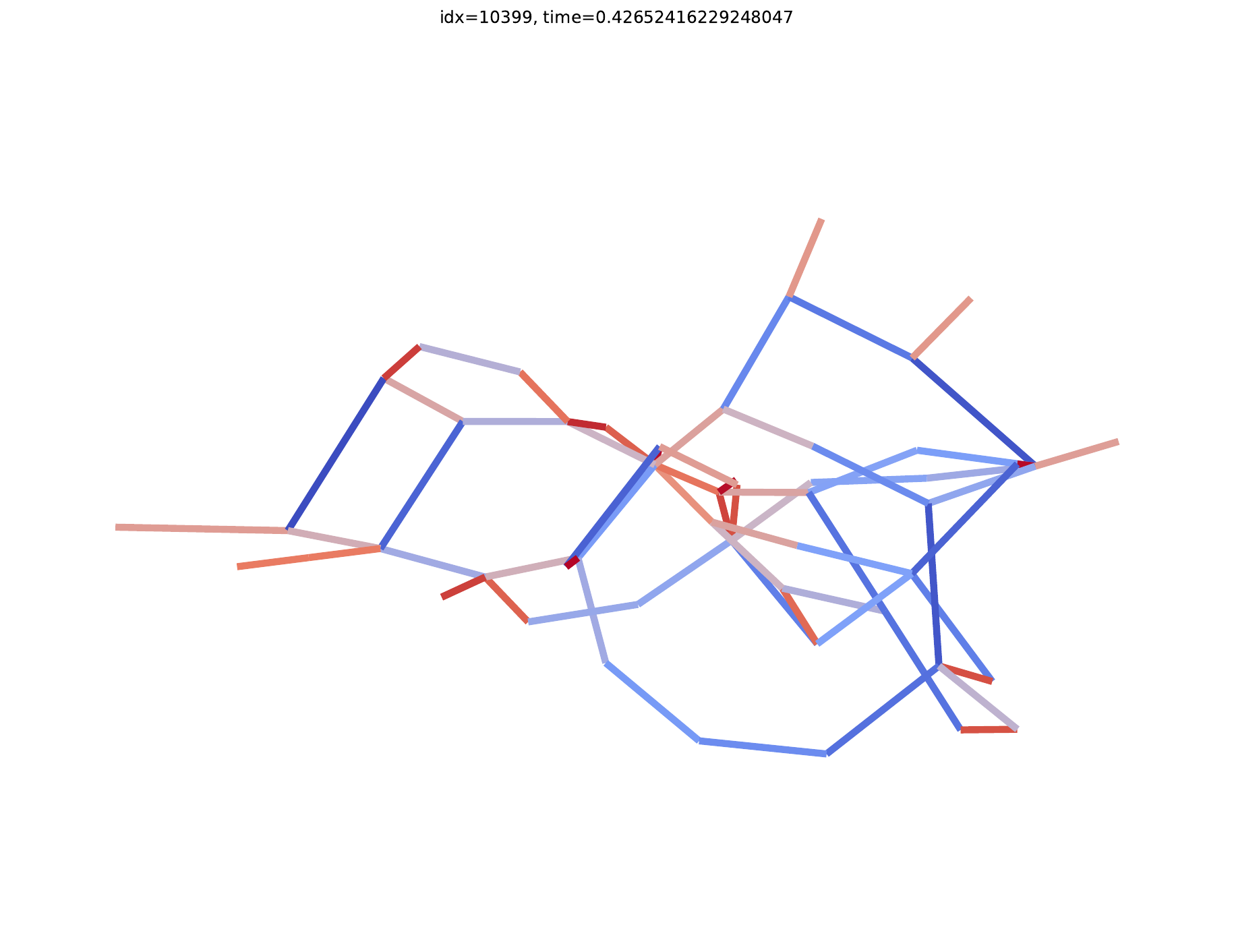} &
\imgcell{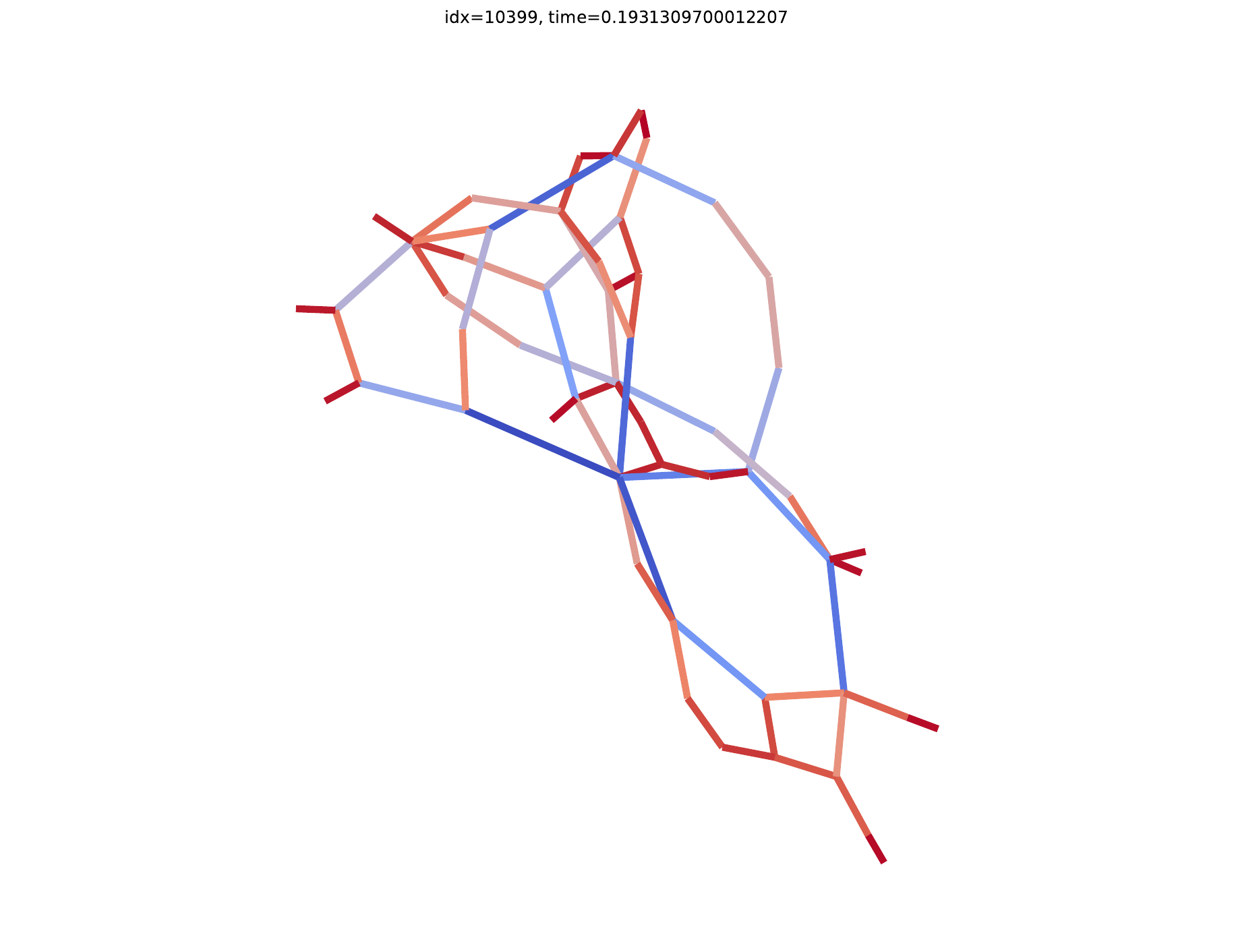} &
\imgcell{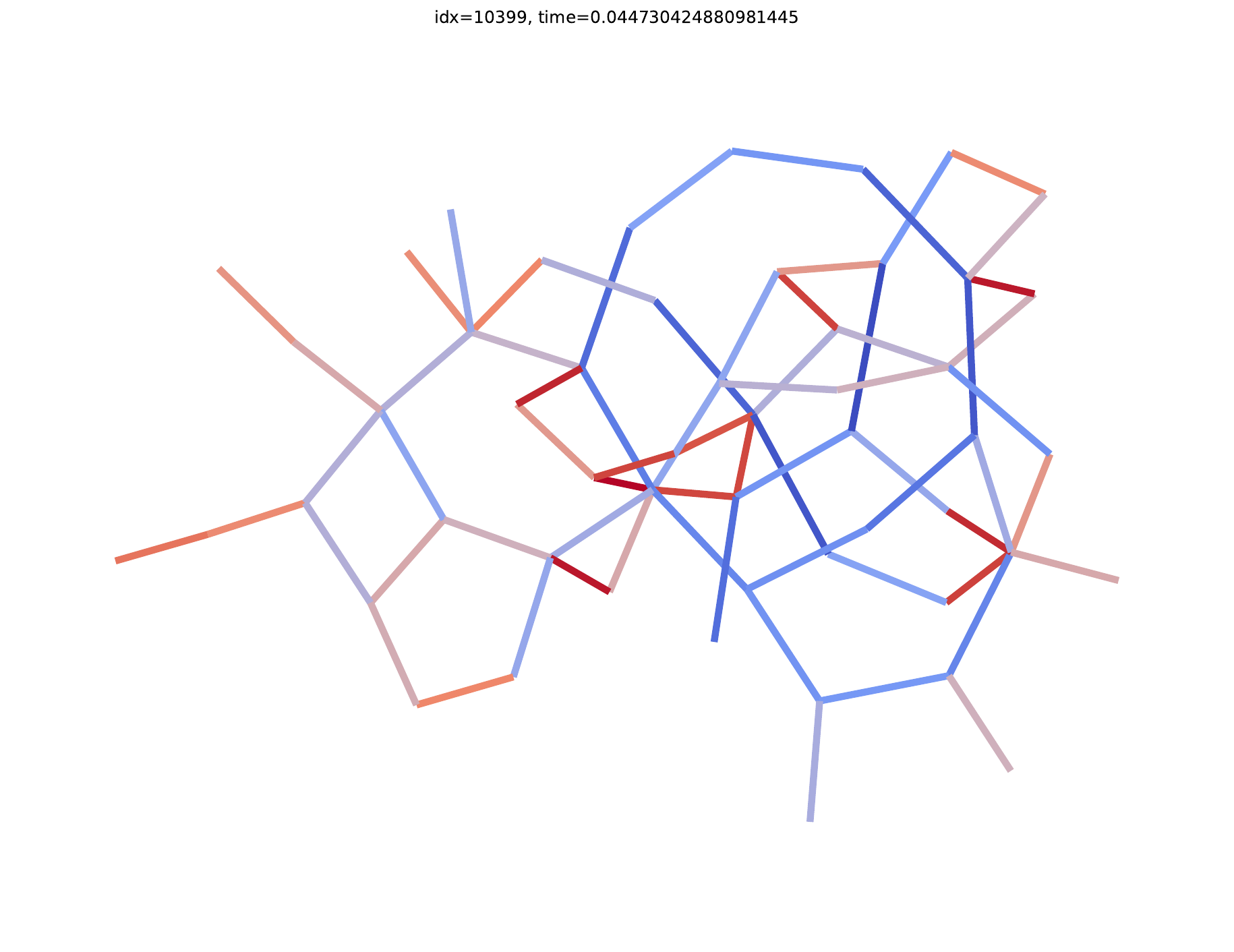} &
\imgcell{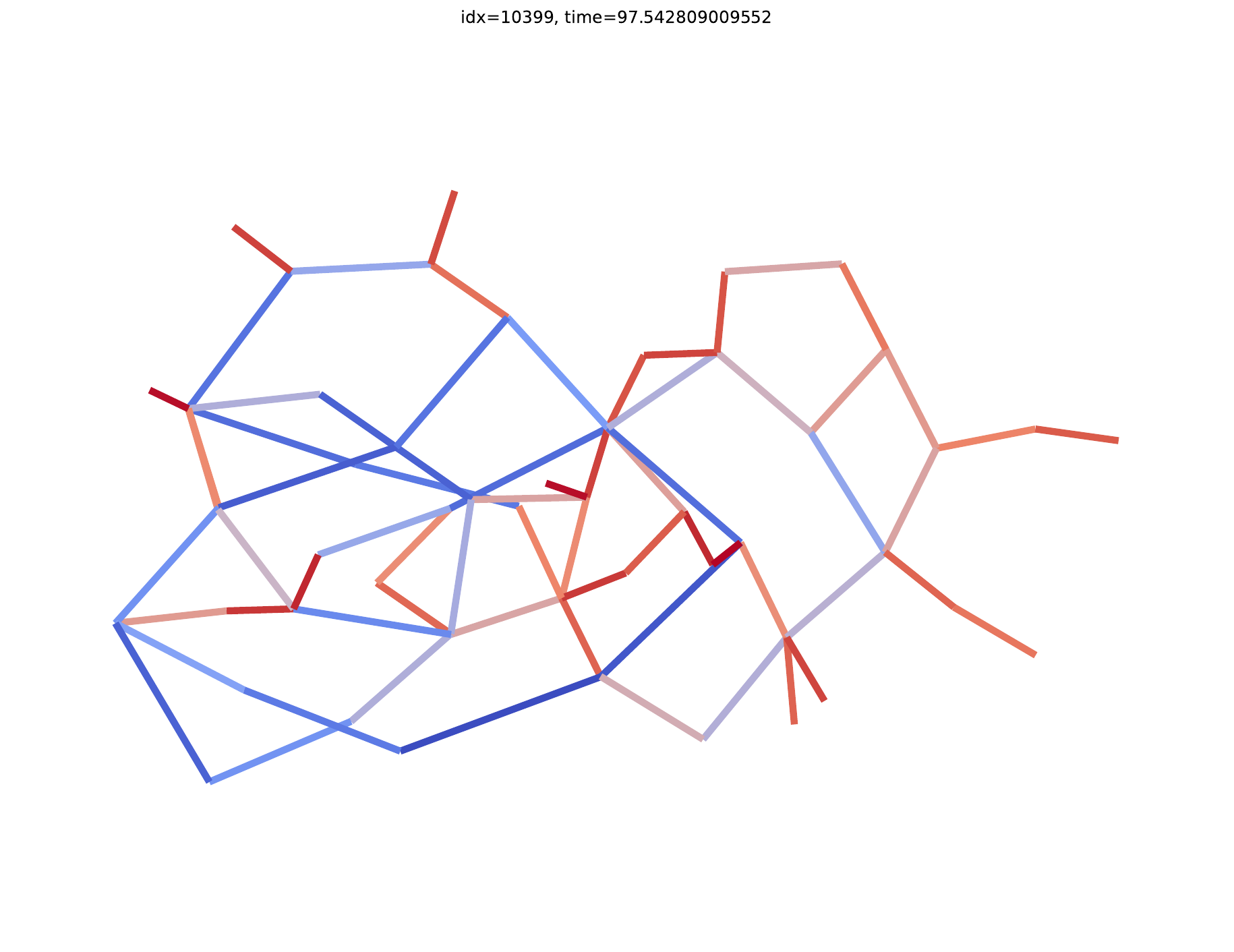} &
\imgcell{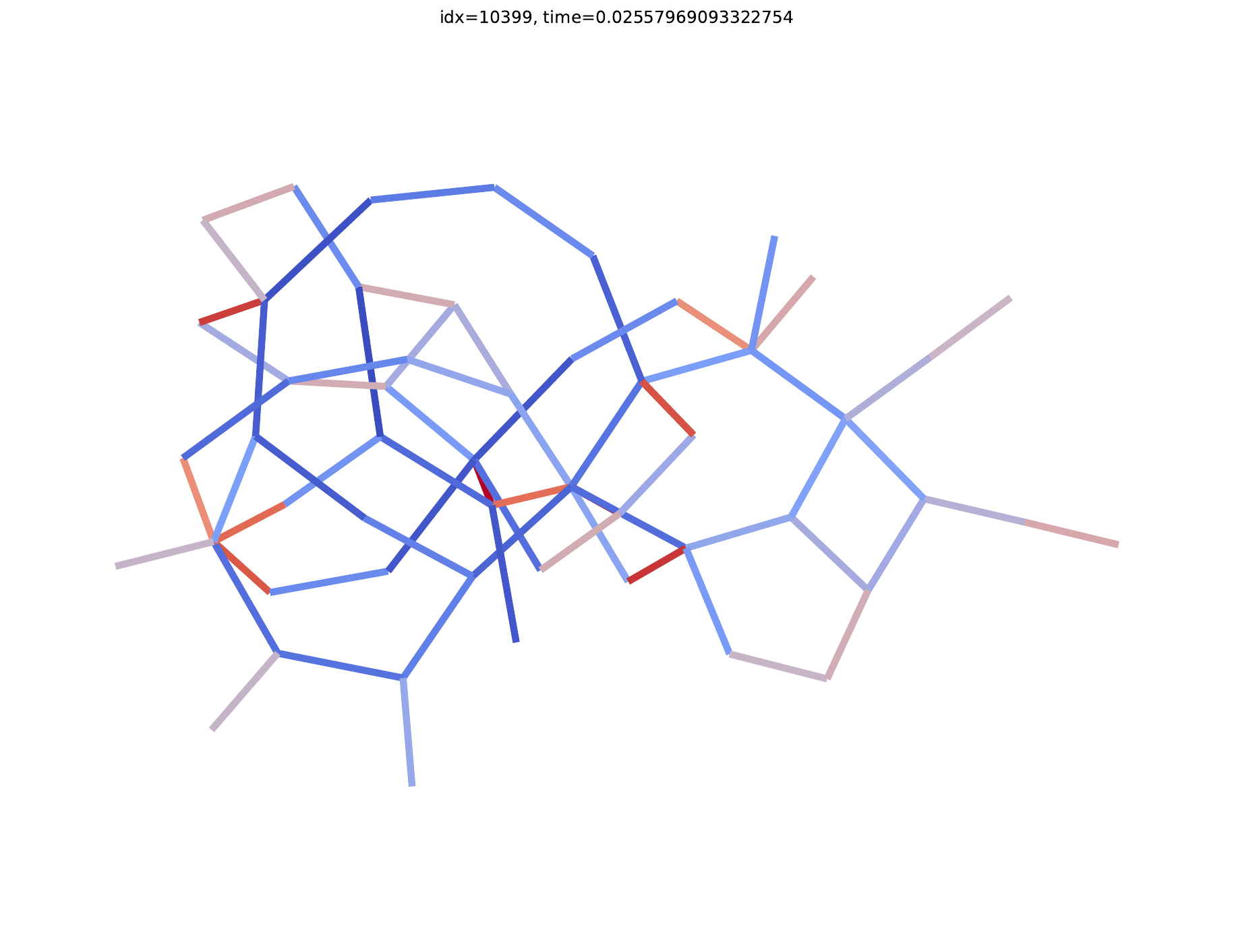} &
\imgcell{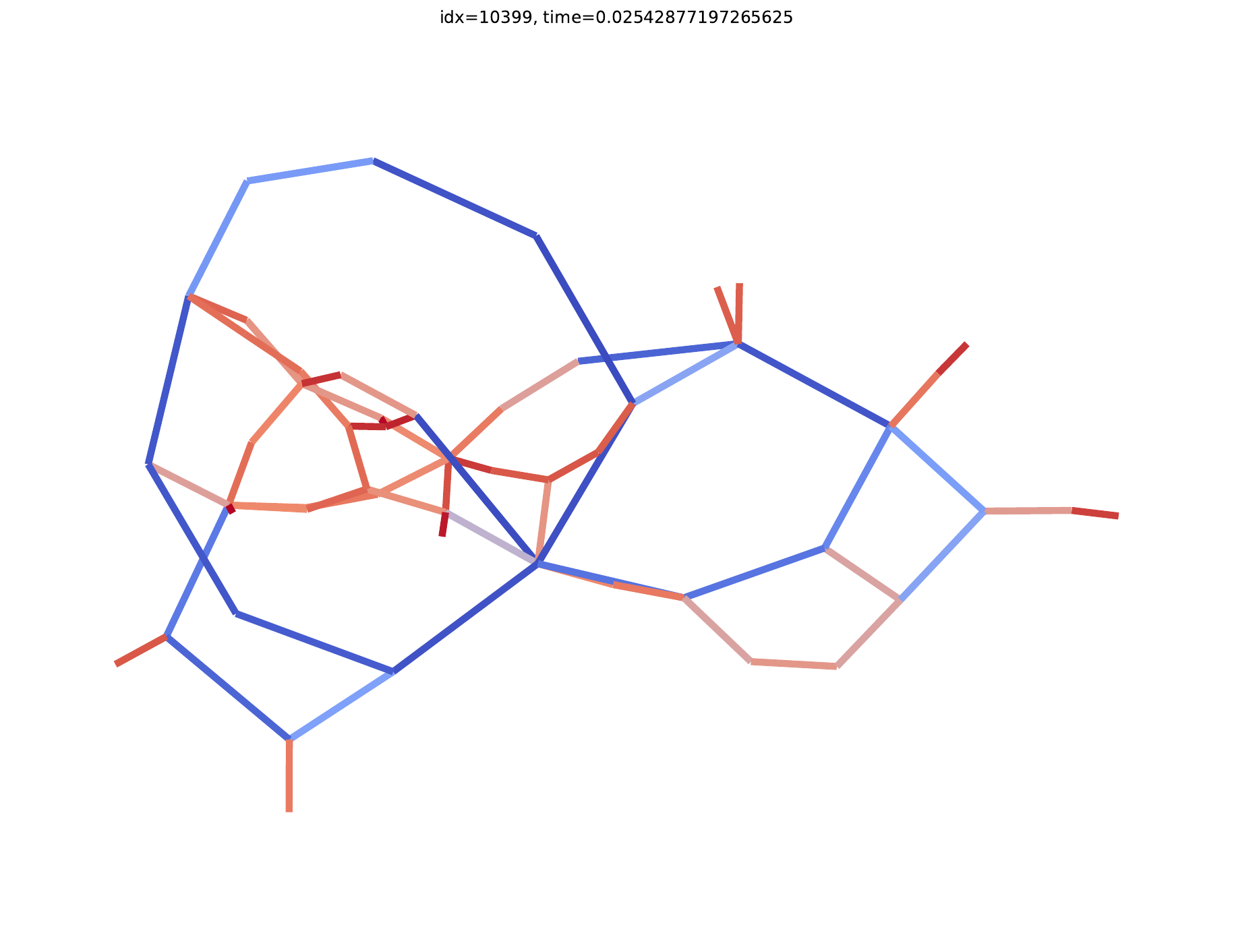} &
\imgcell{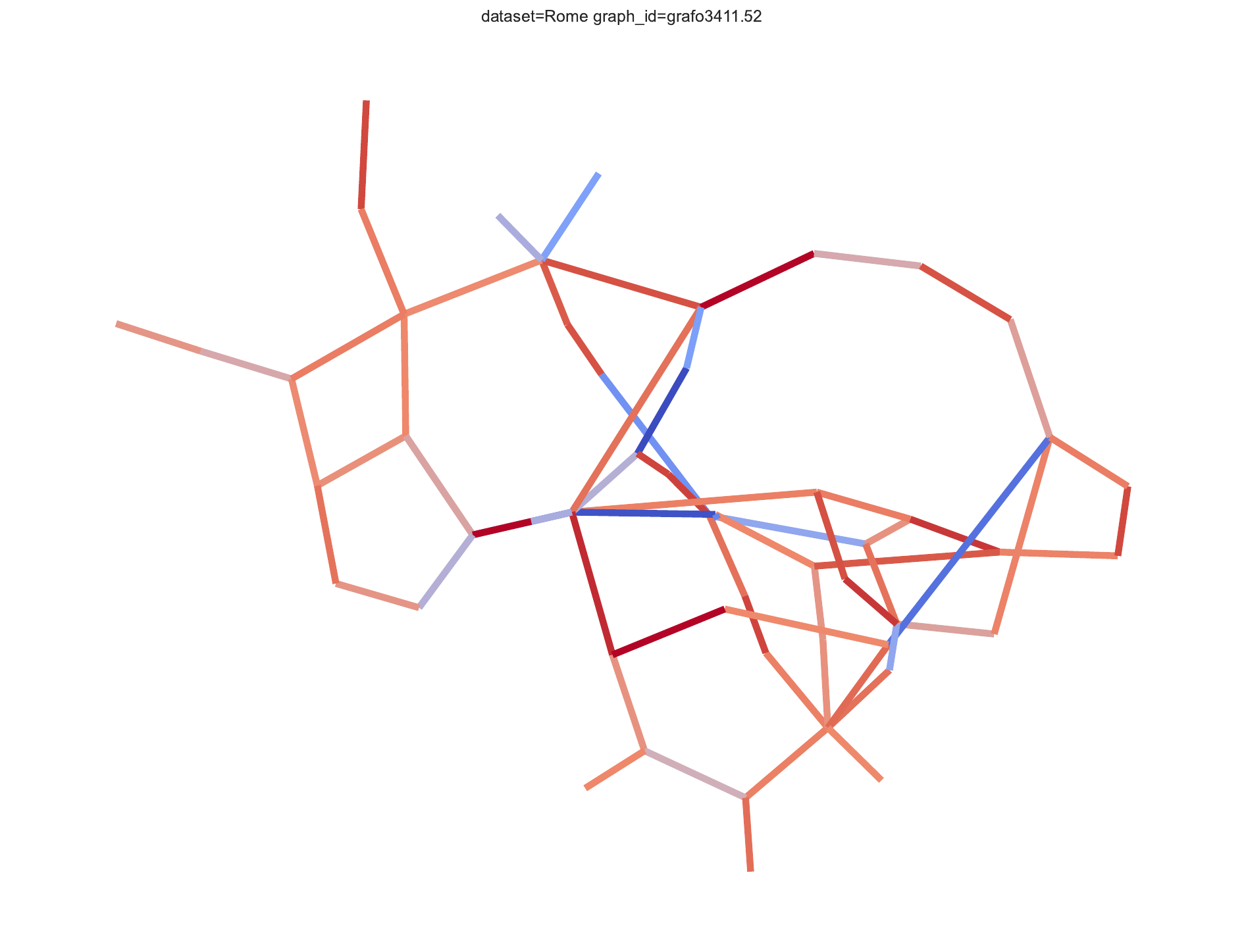} &
\imgcell{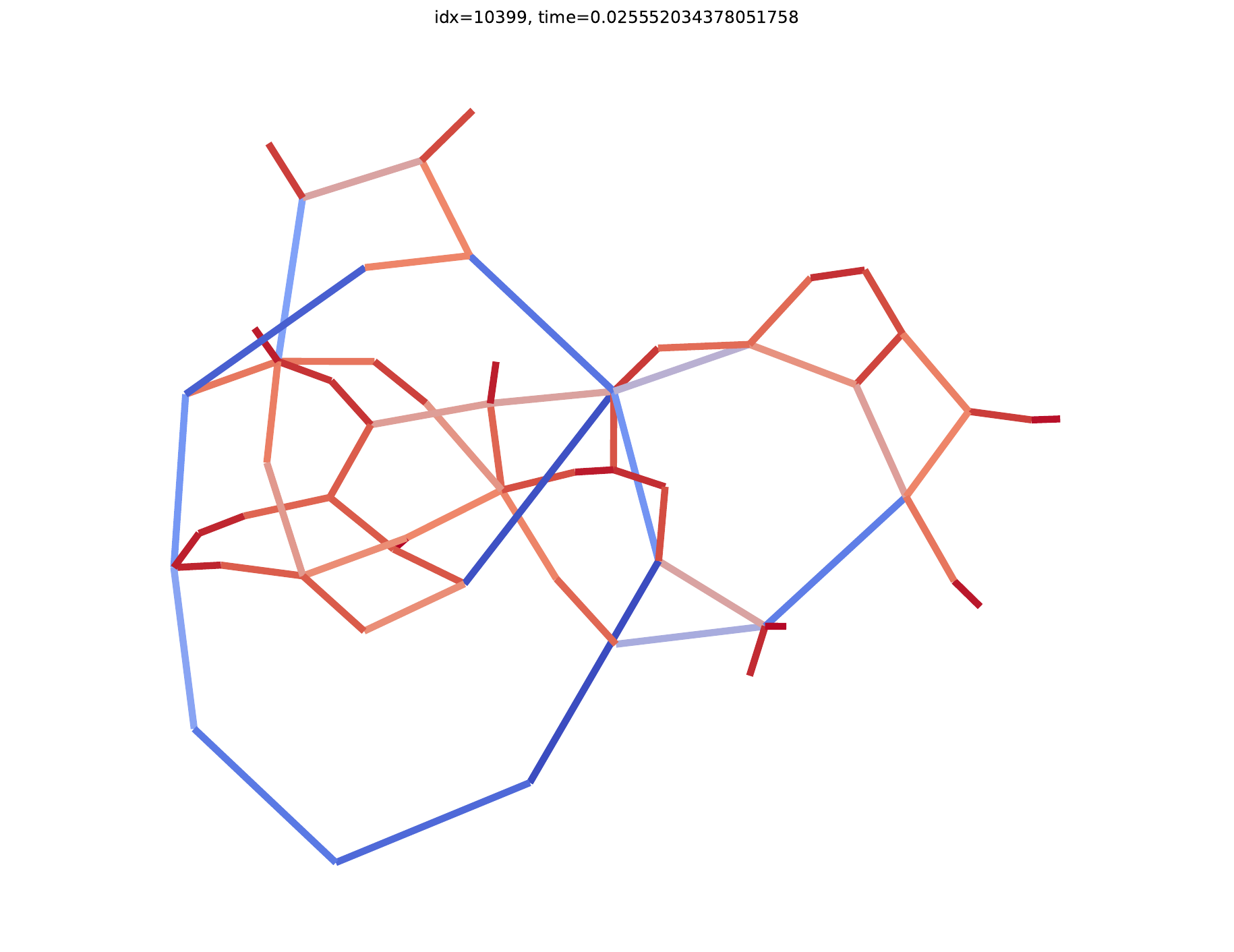} &
\imgcell{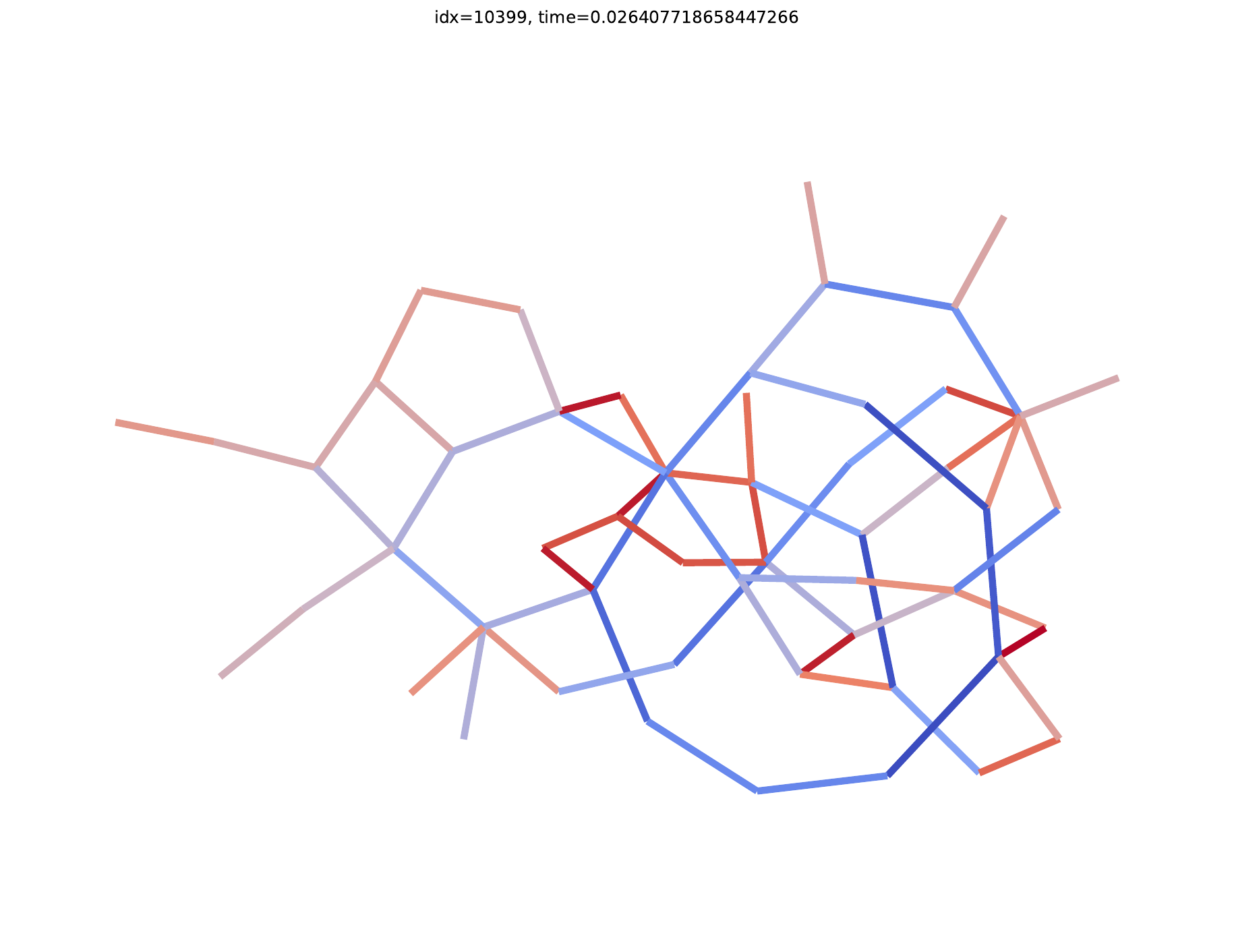} &
\imgcell{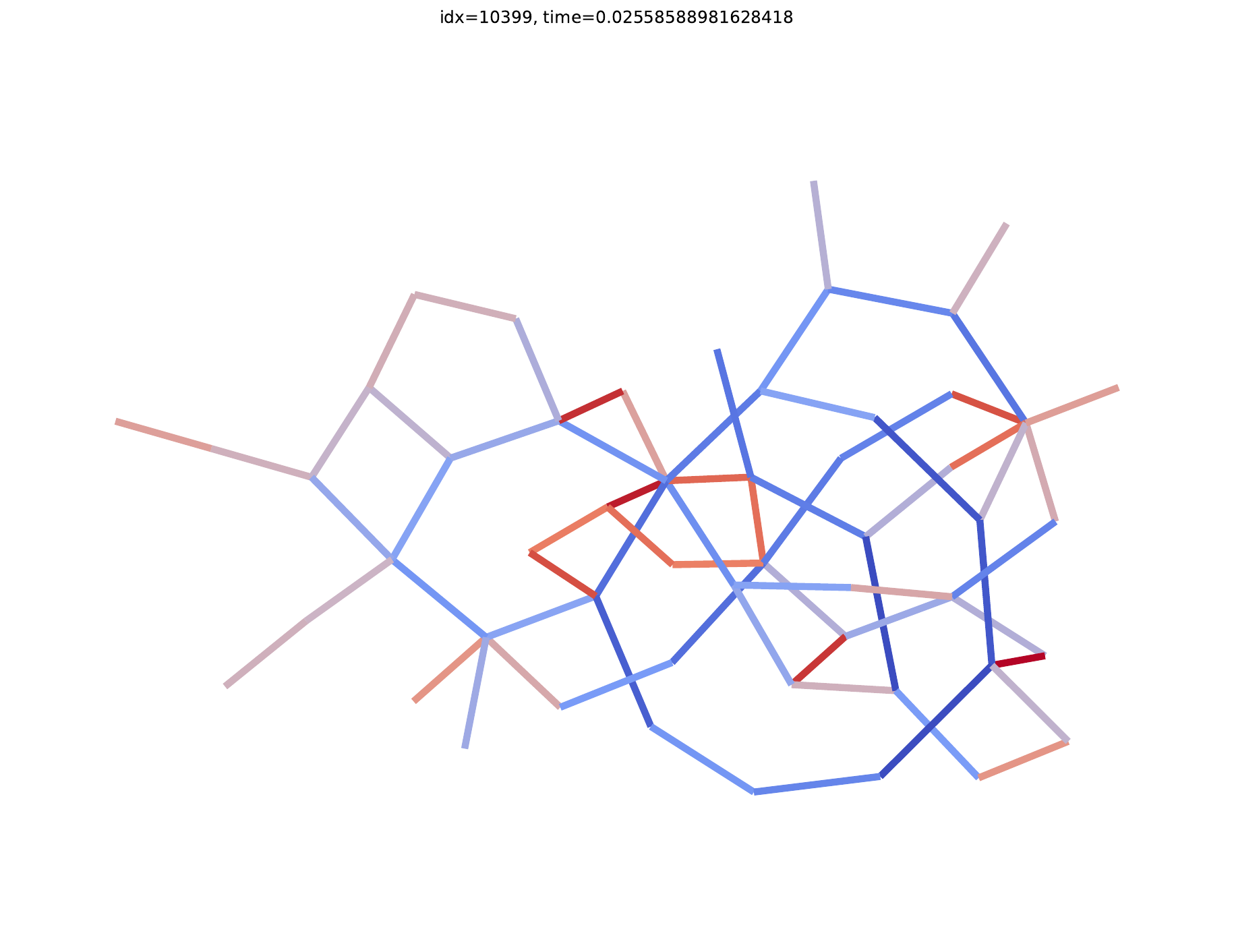} &
\imgcell{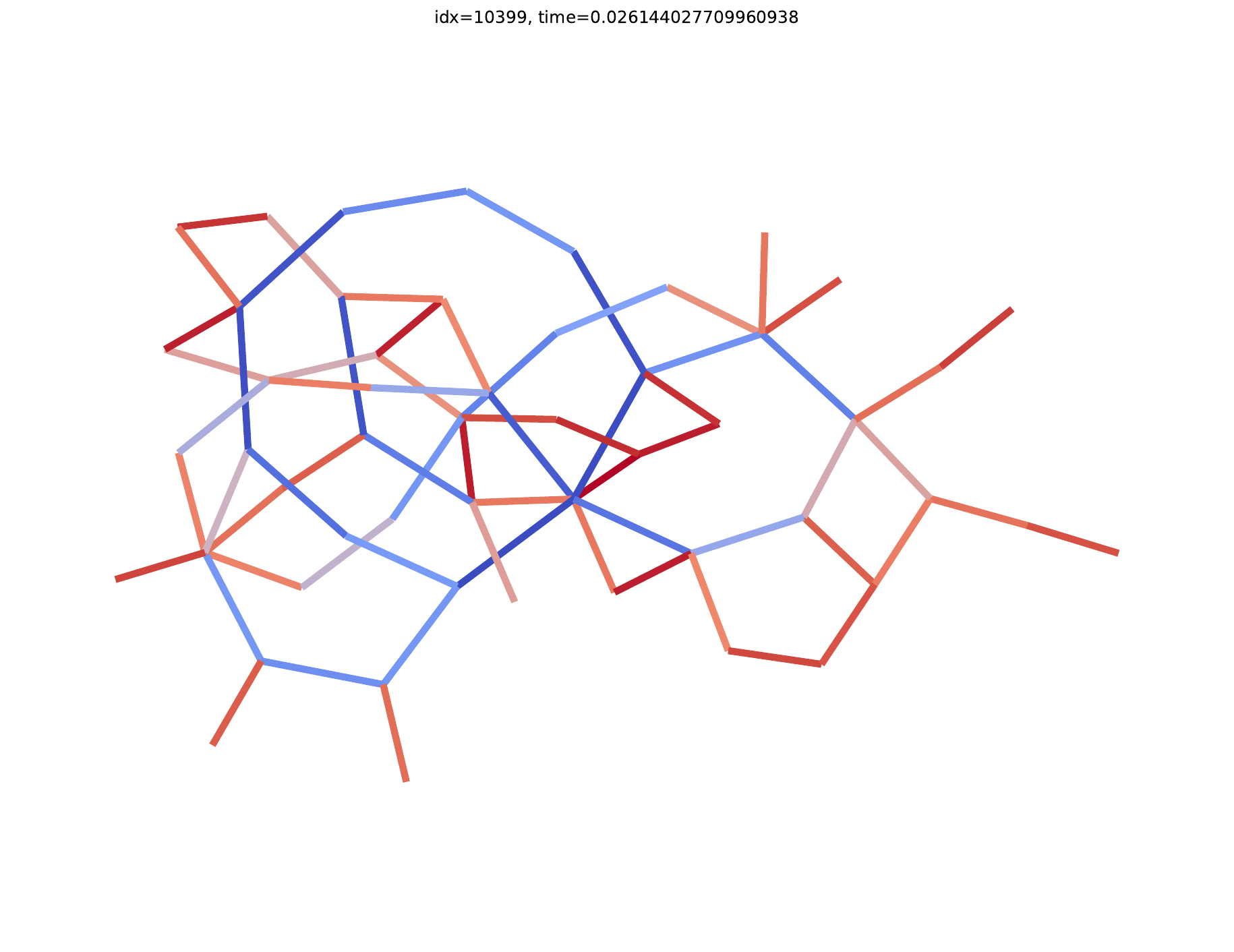} \\

&
t = 0.00s &
t = 0.43s &
t = 0.19s &
t = 0.04s &
t = 97.54s &
t = 0.03s &
t = 0.03s &
t = 0.03s &
t = 0.03s &
t = 0.03s &
t = 0.03s &
t = 0.03s \\

\makecell{\bfseries rgrafo1810.30\\N = 14\\M = 15} &
\imgcell{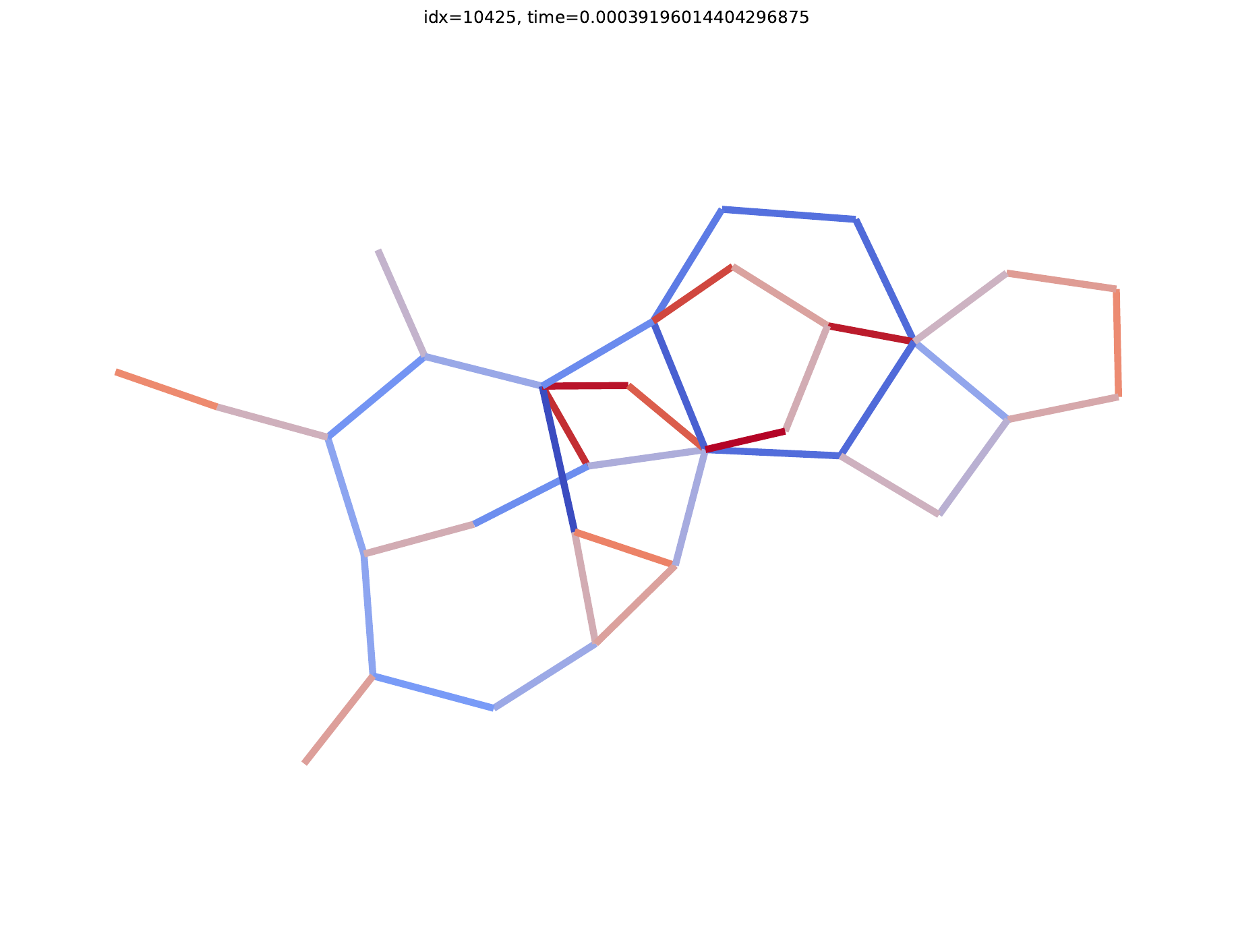} &
\imgcell{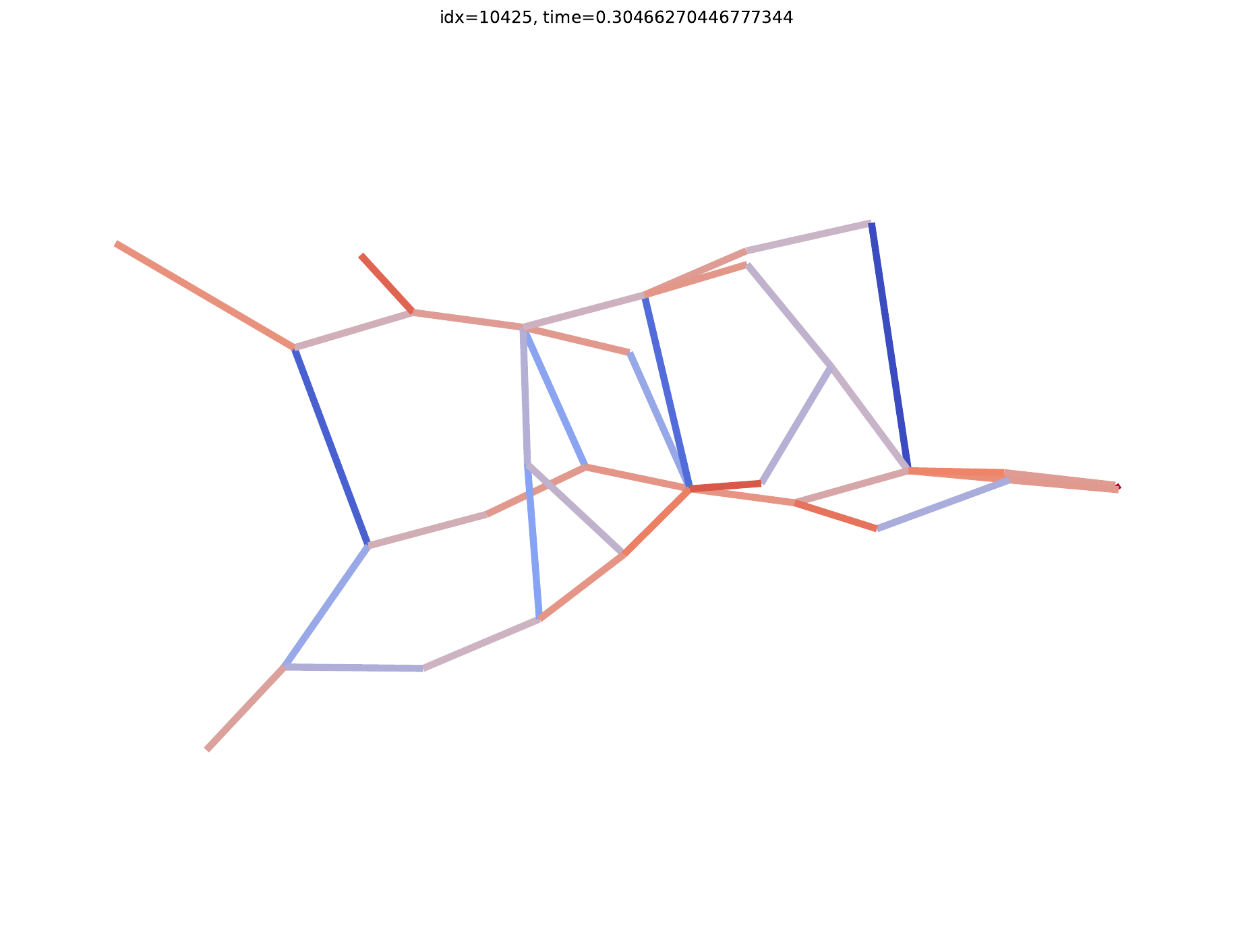} &
\imgcell{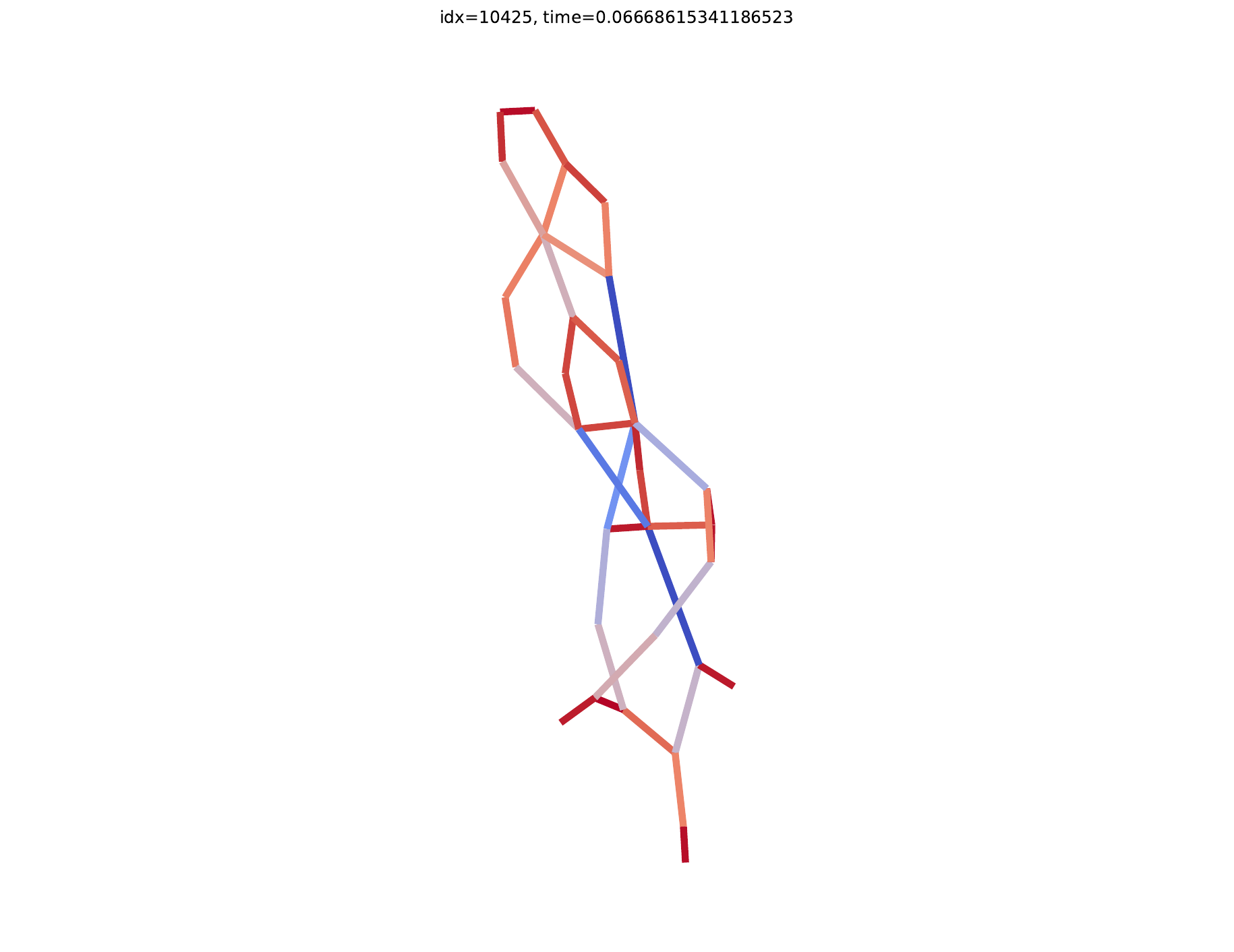} &
\imgcell{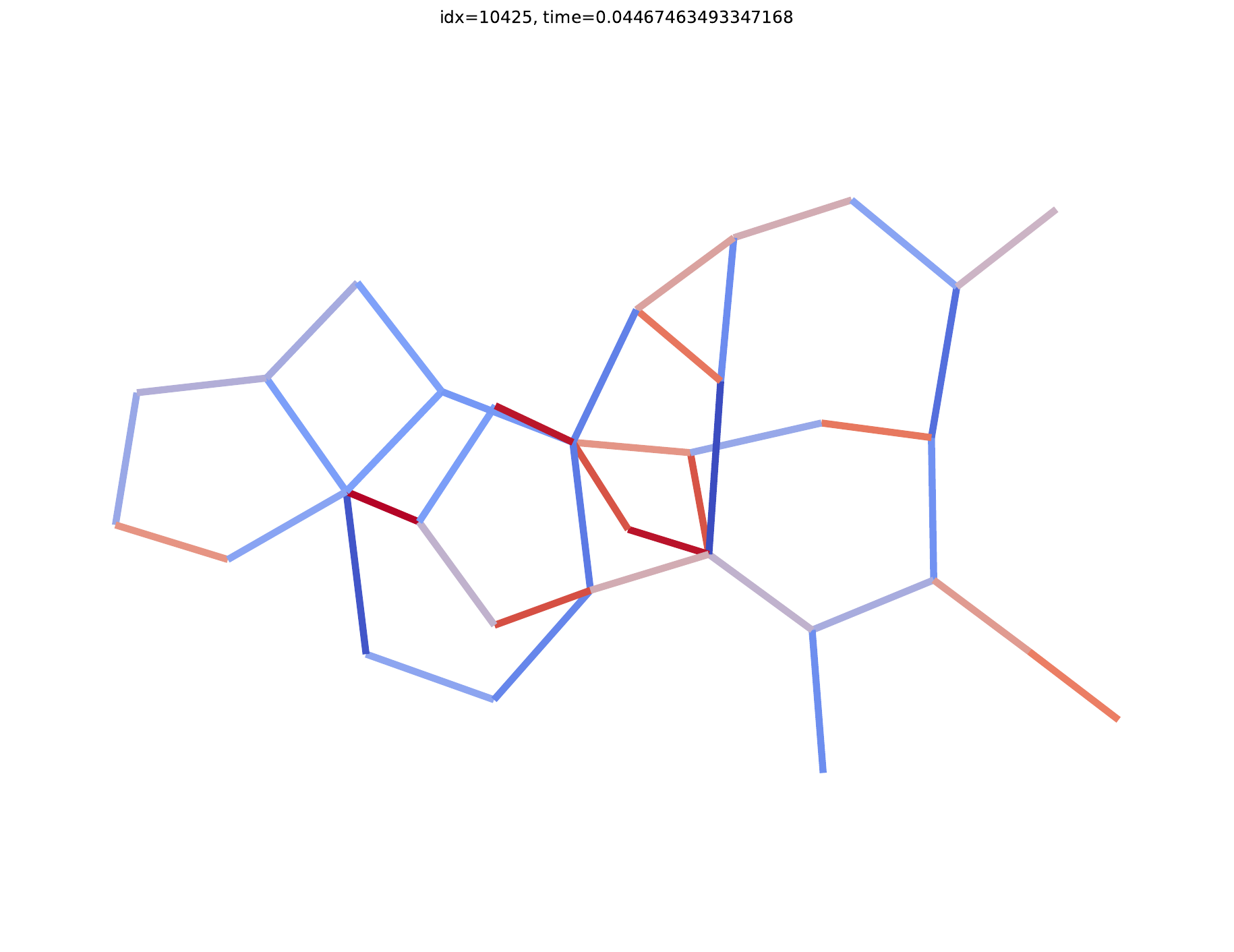} &
\imgcell{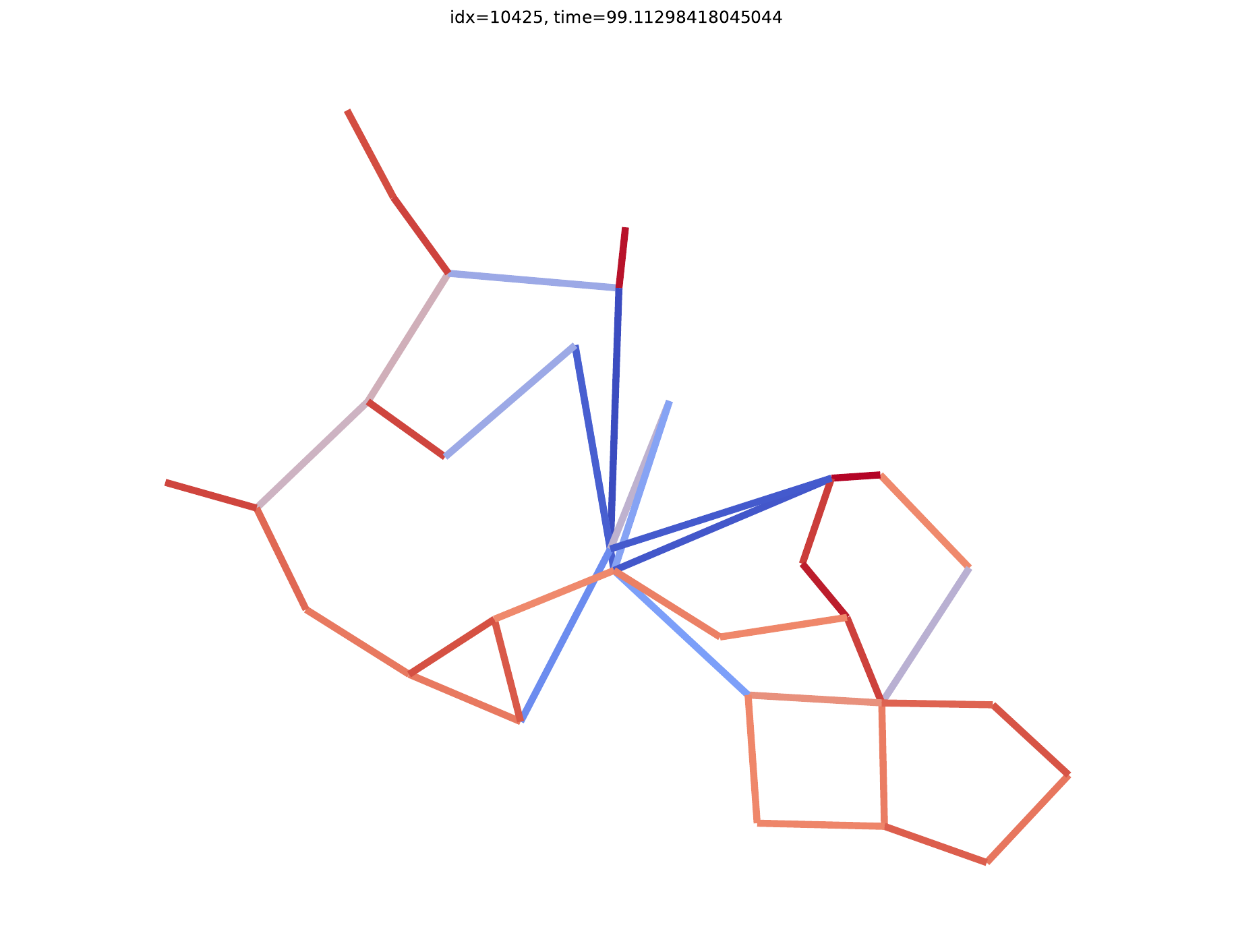} &
\imgcell{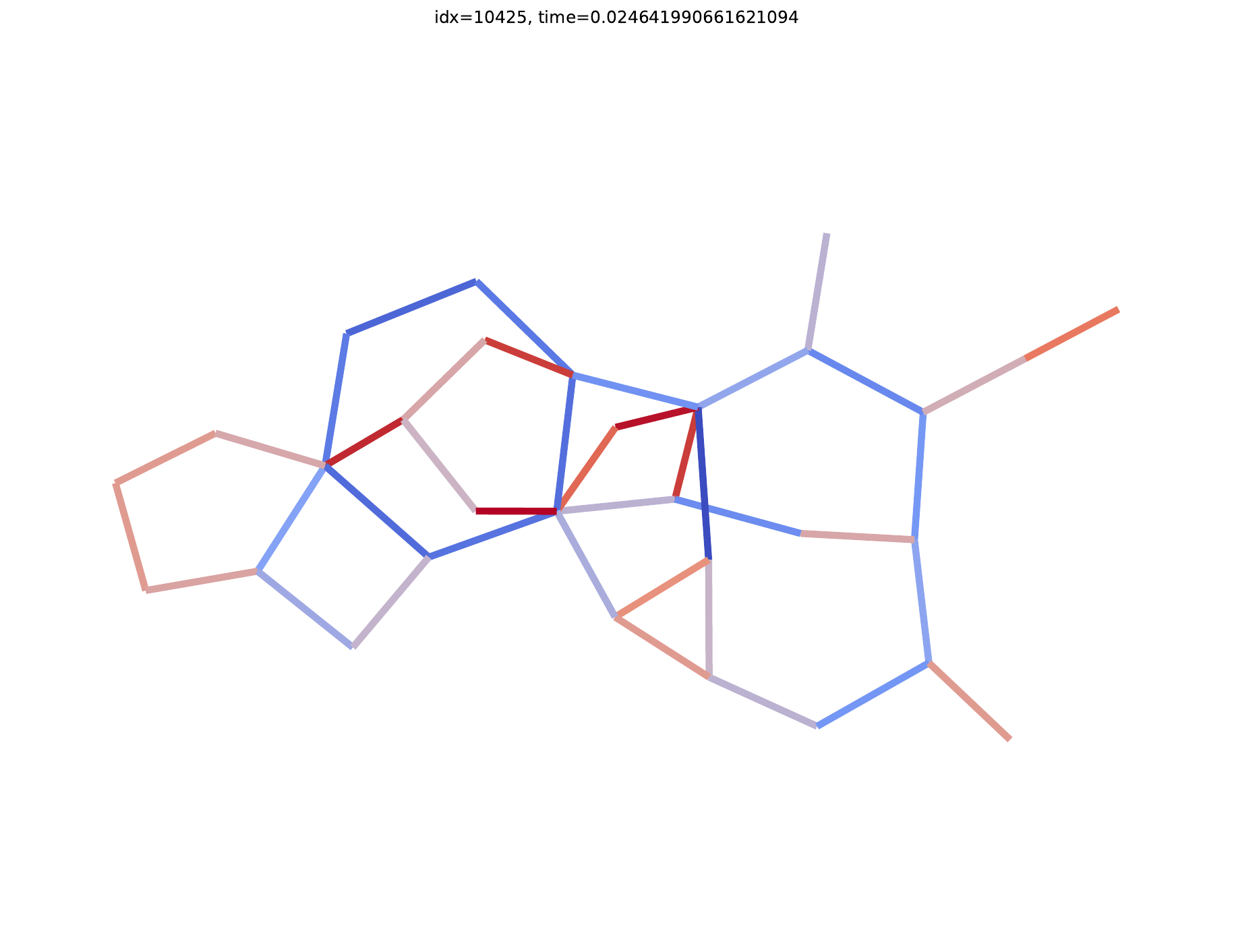} &
\imgcell{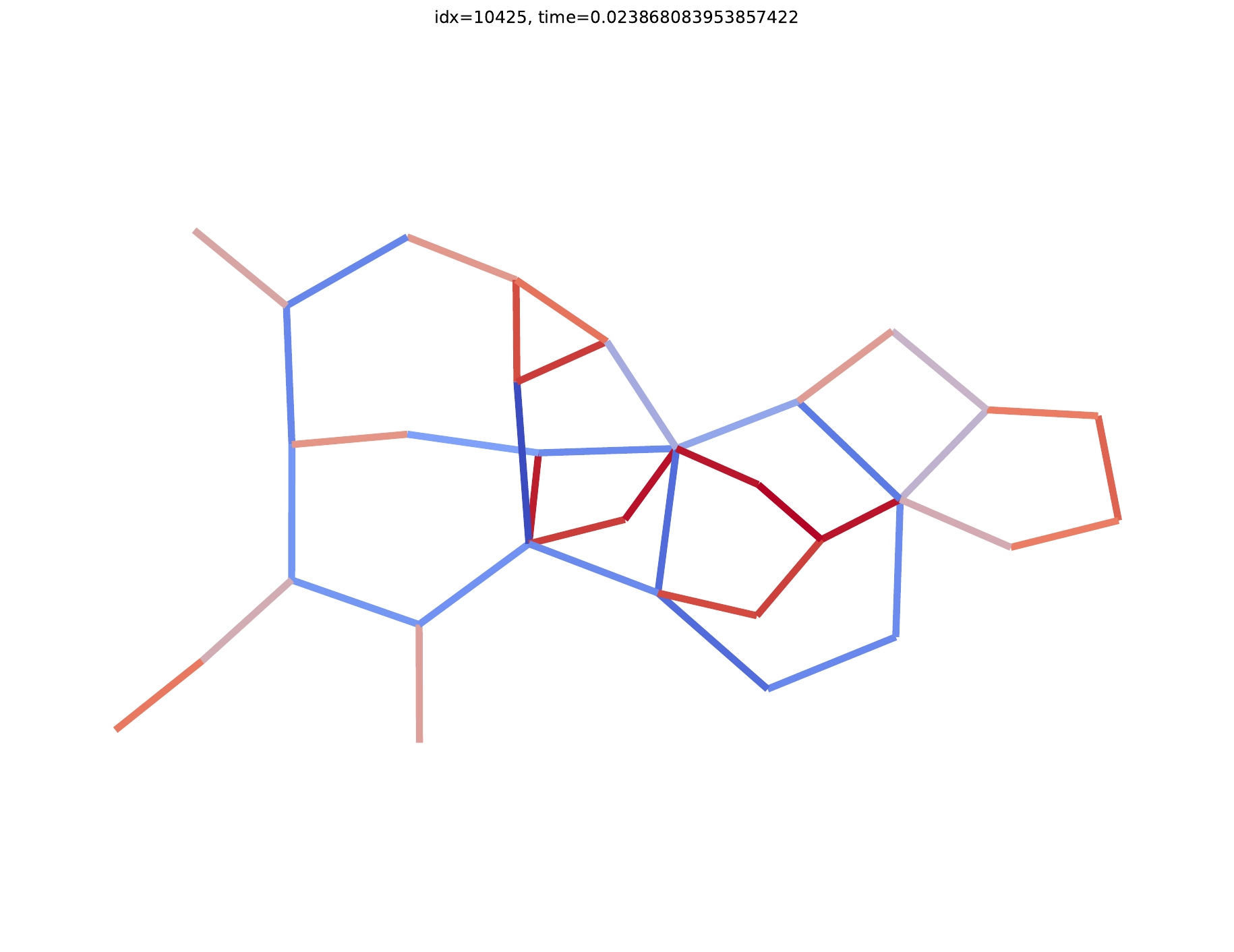} &
\imgcell{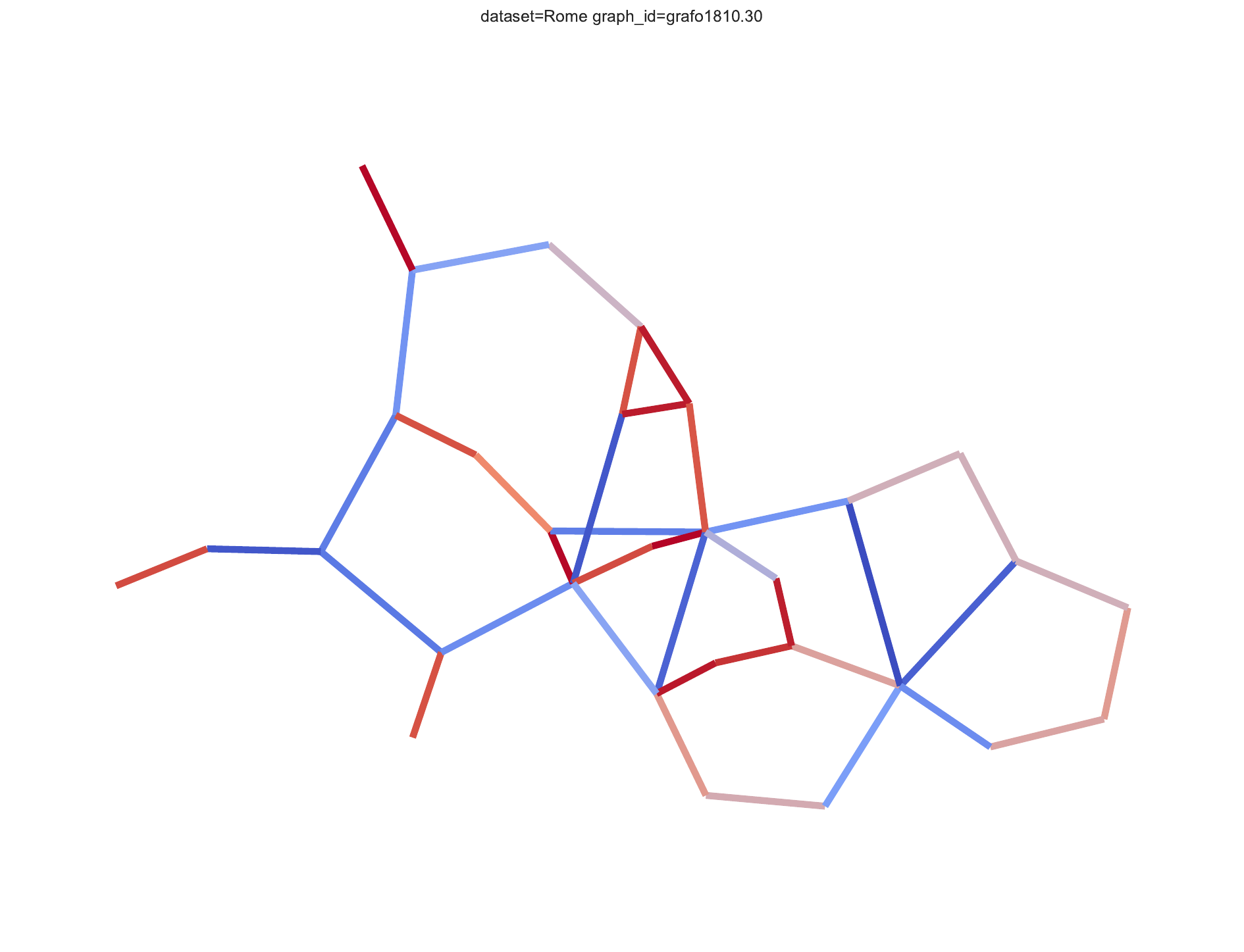} &
\imgcell{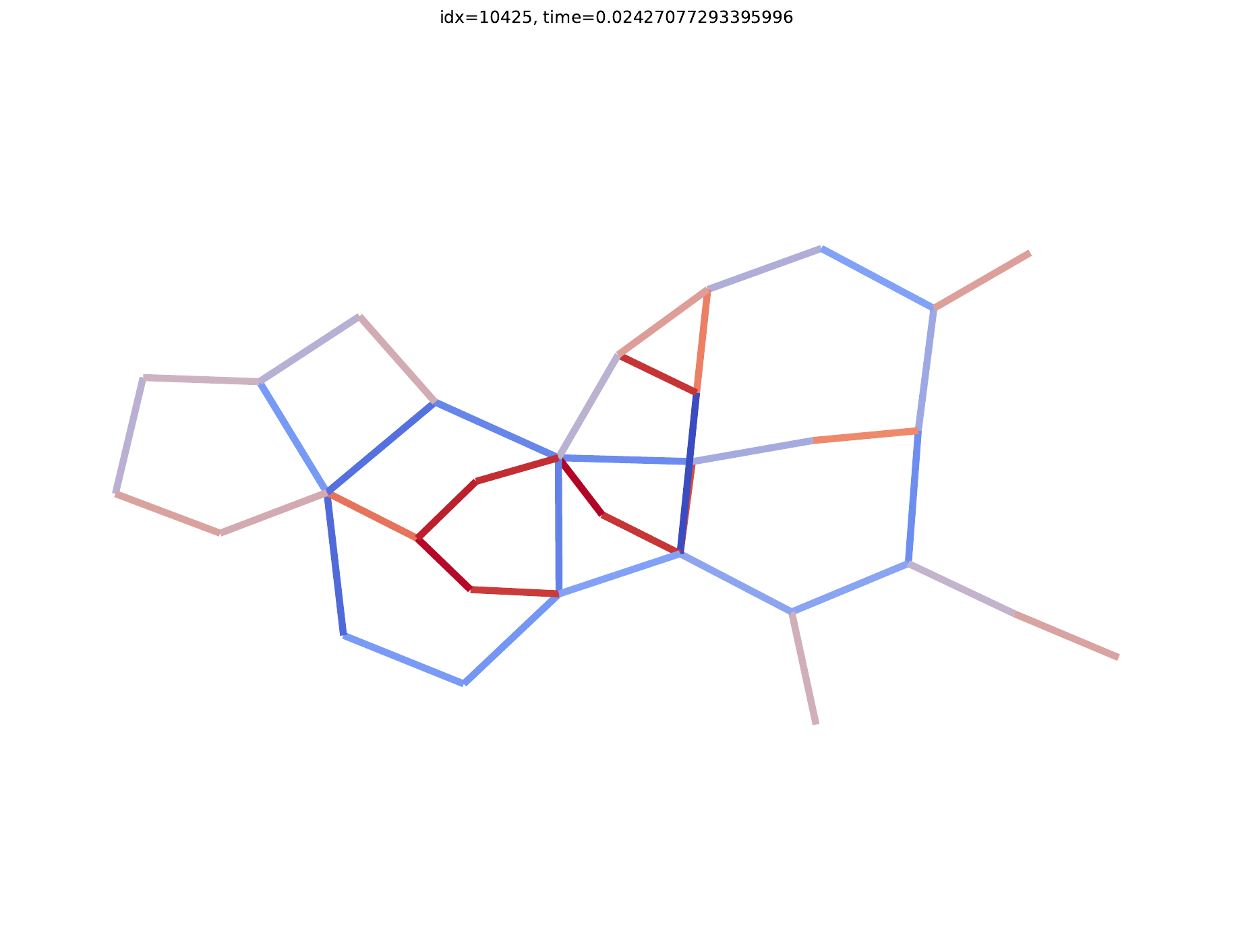} &
\imgcell{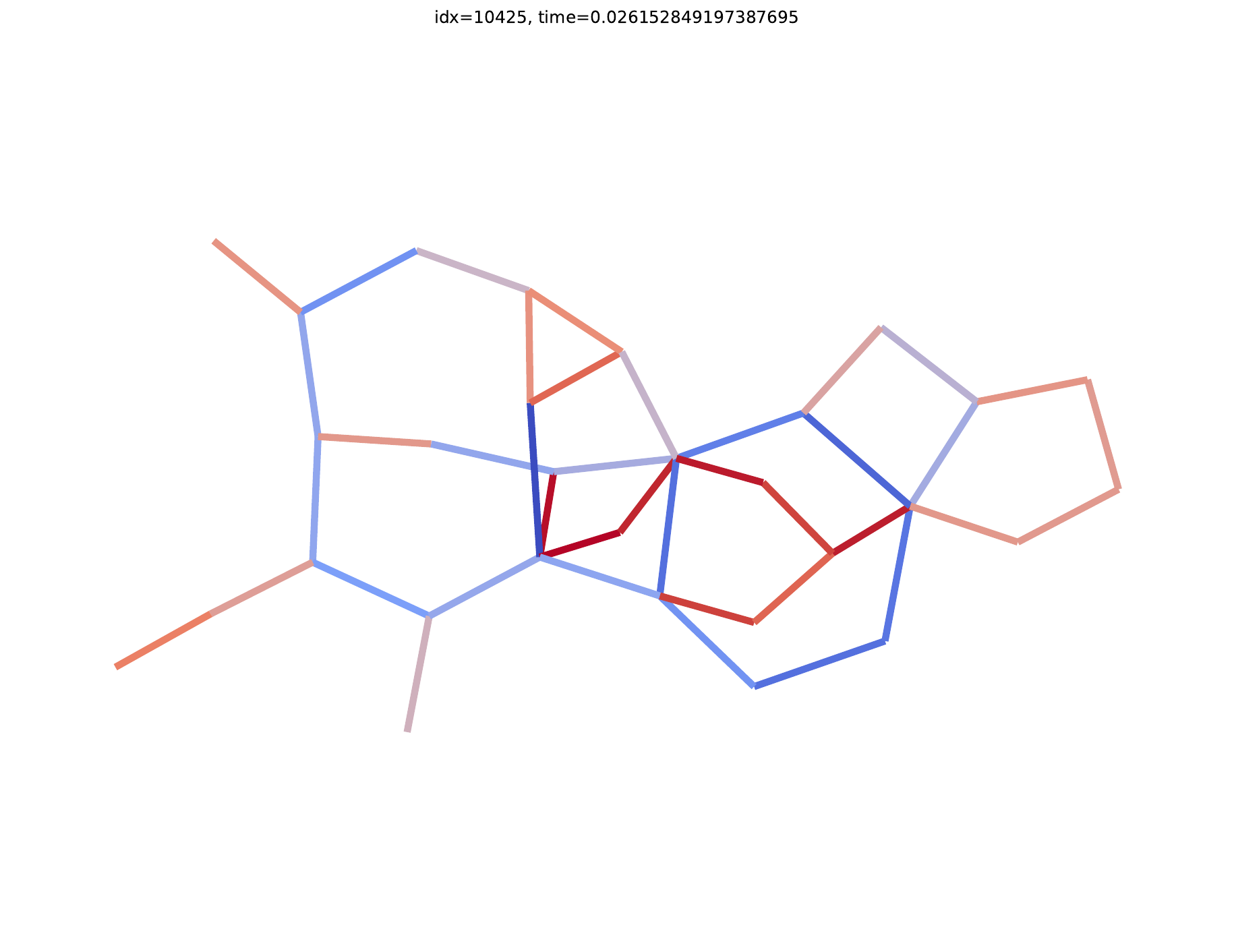} &
\imgcell{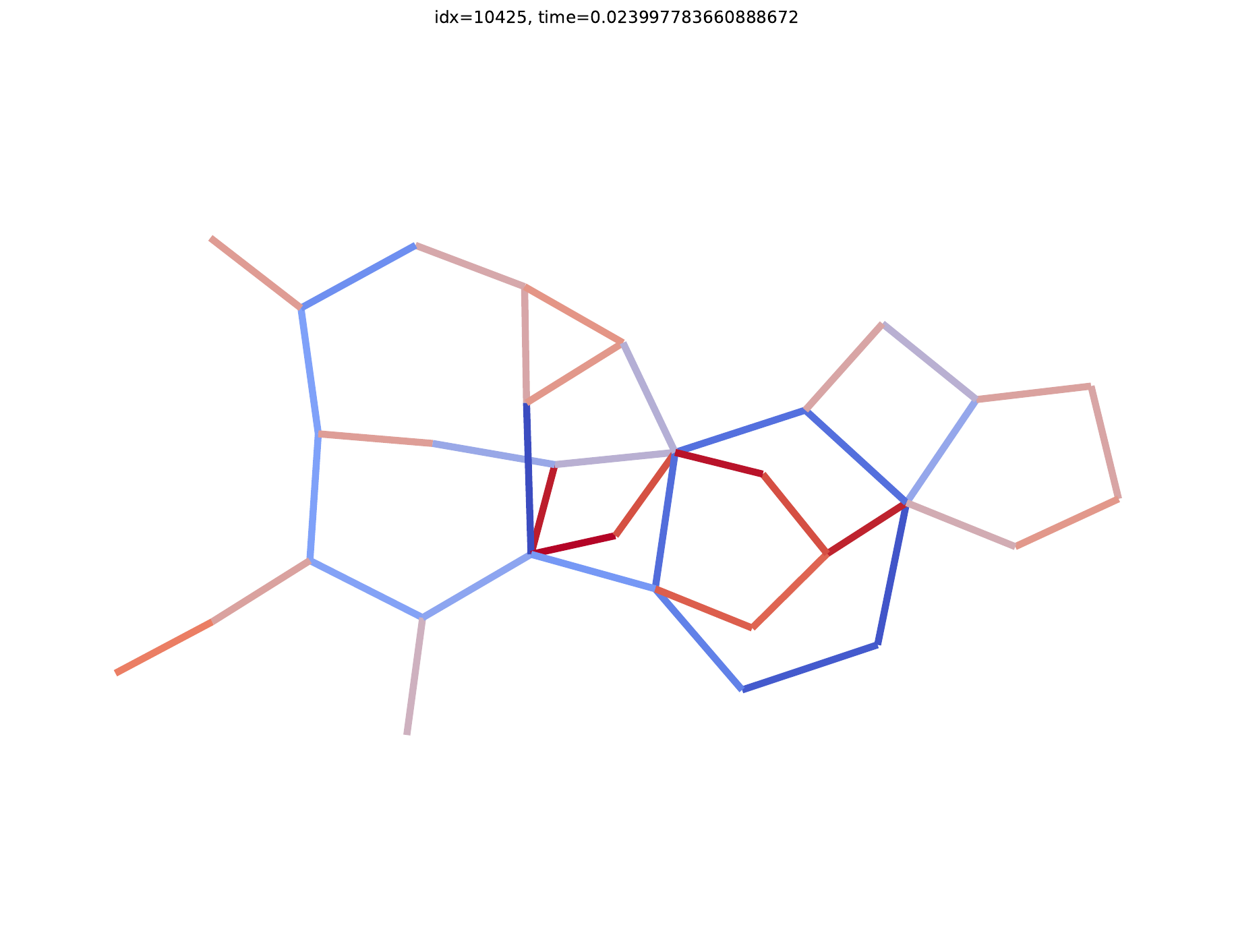} &
\imgcell{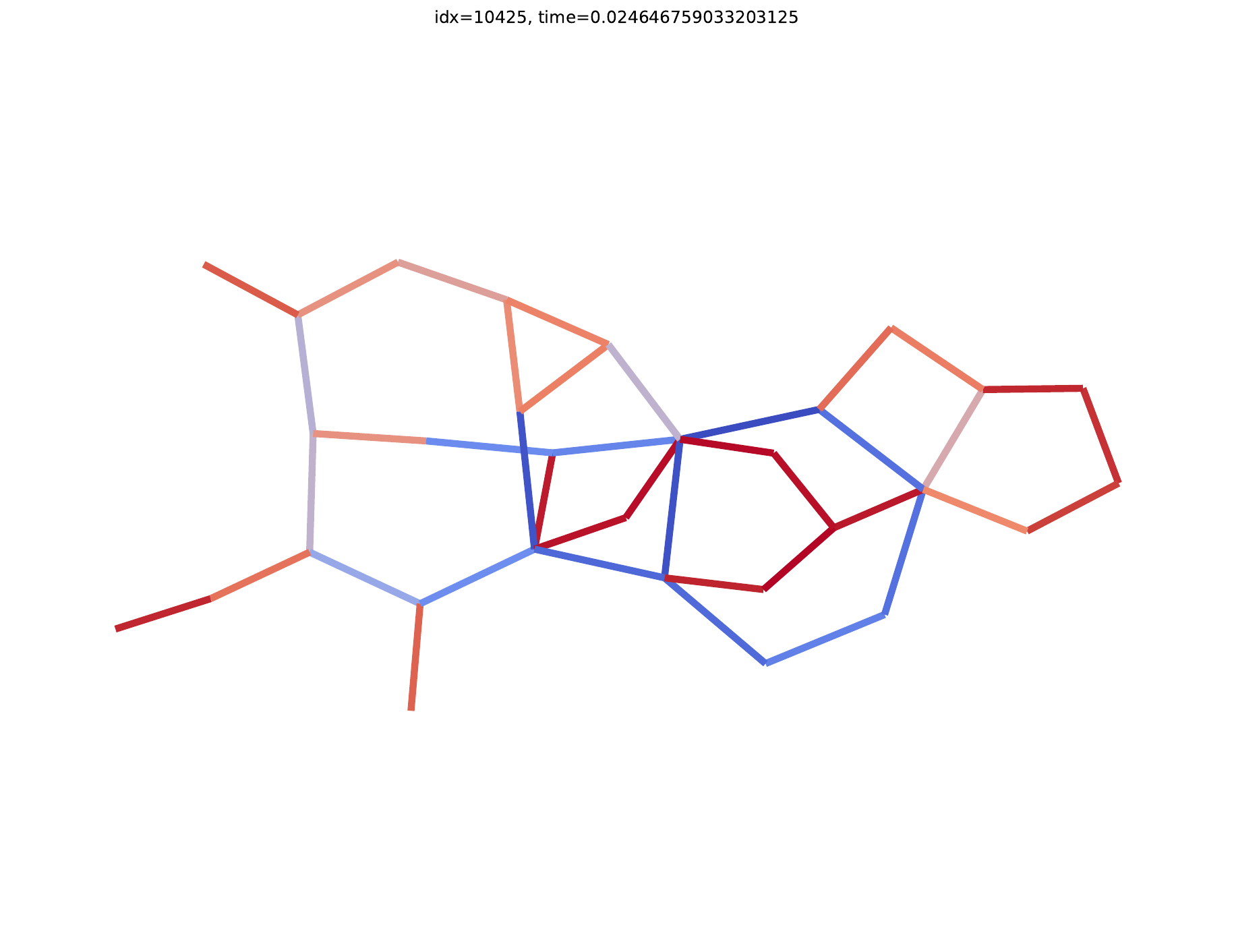} \\

&
t = 0.00s &
t = 0.30s &
t = 0.07s &
t = 0.04s &
t = 99.11s &
t = 0.02s &
t = 0.02s &
t = 0.03s &
t = 0.02s &
t = 0.03s &
t = 0.02s &
t = 0.02s \\

\makecell{\bfseries grafo7515.81\\N = 100\\M = 141} &
\imgcell{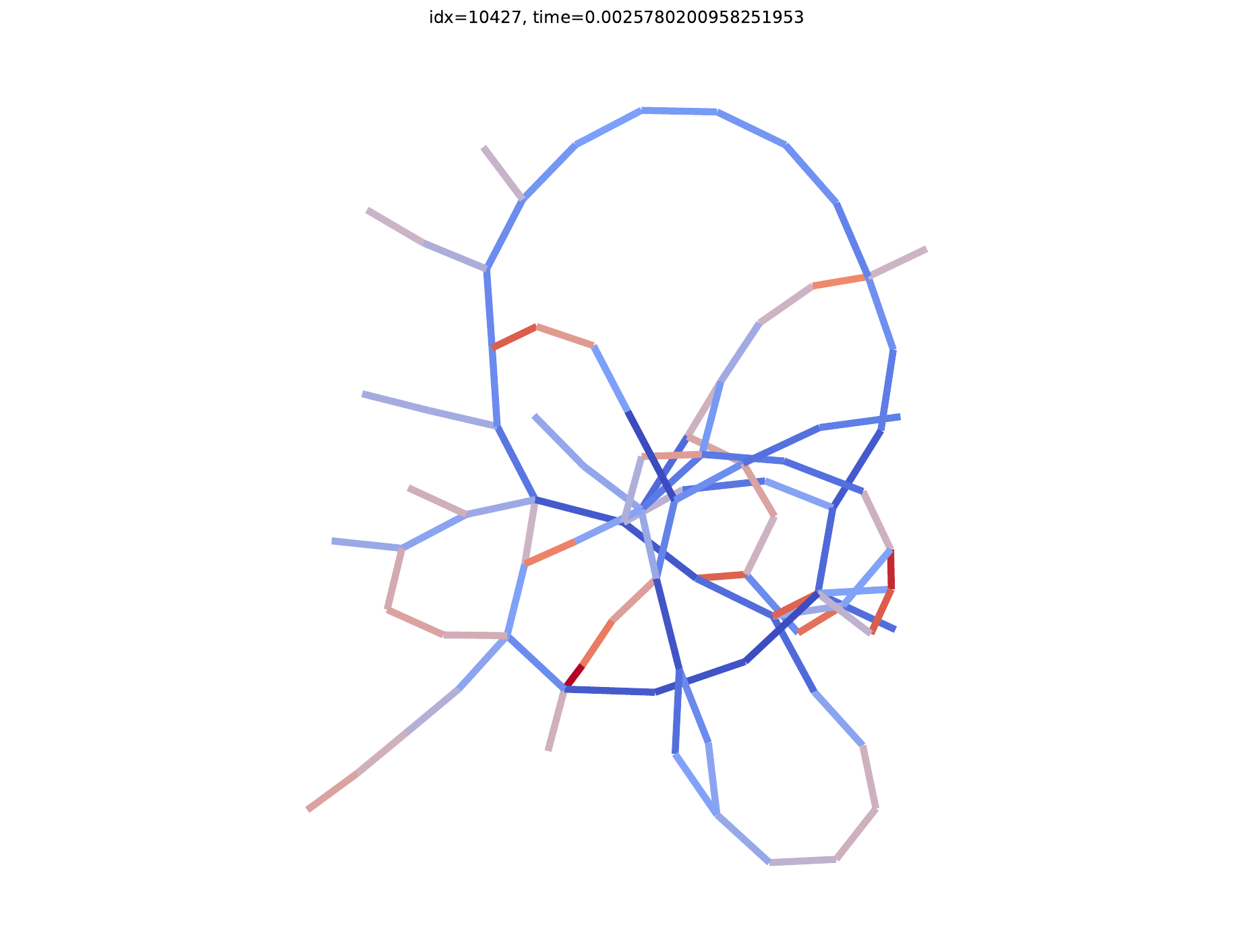} &
\imgcell{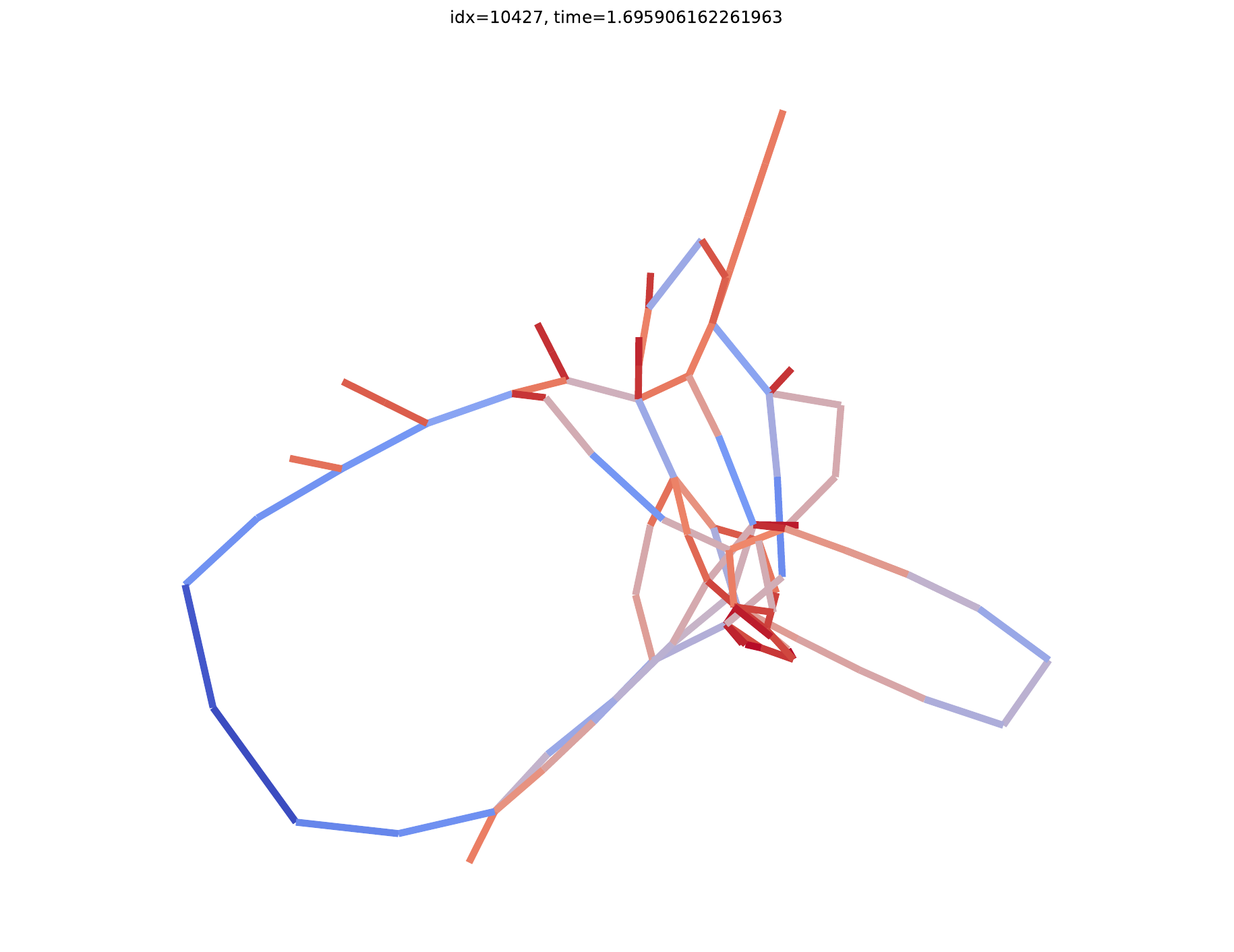} &
\imgcell{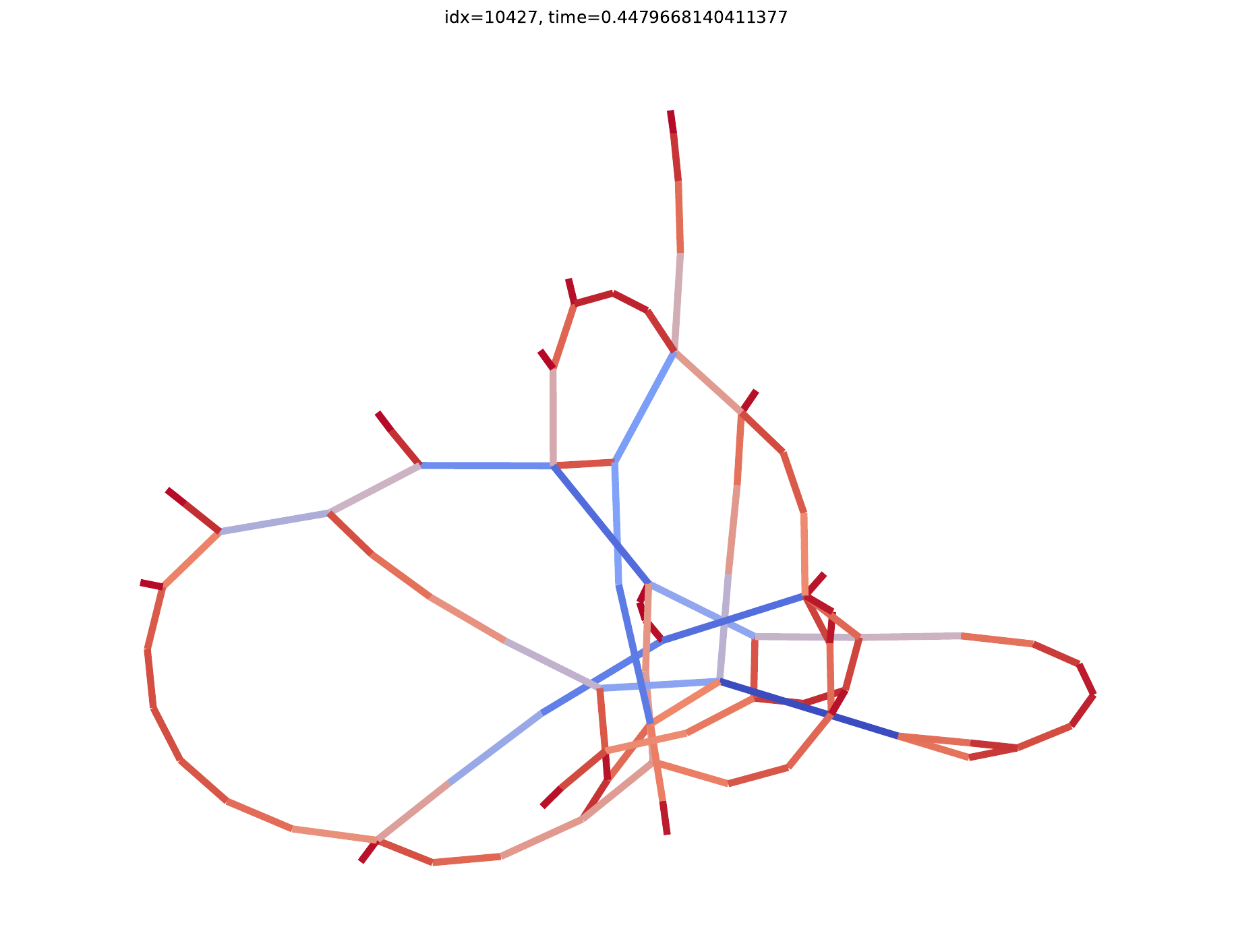} &
\imgcell{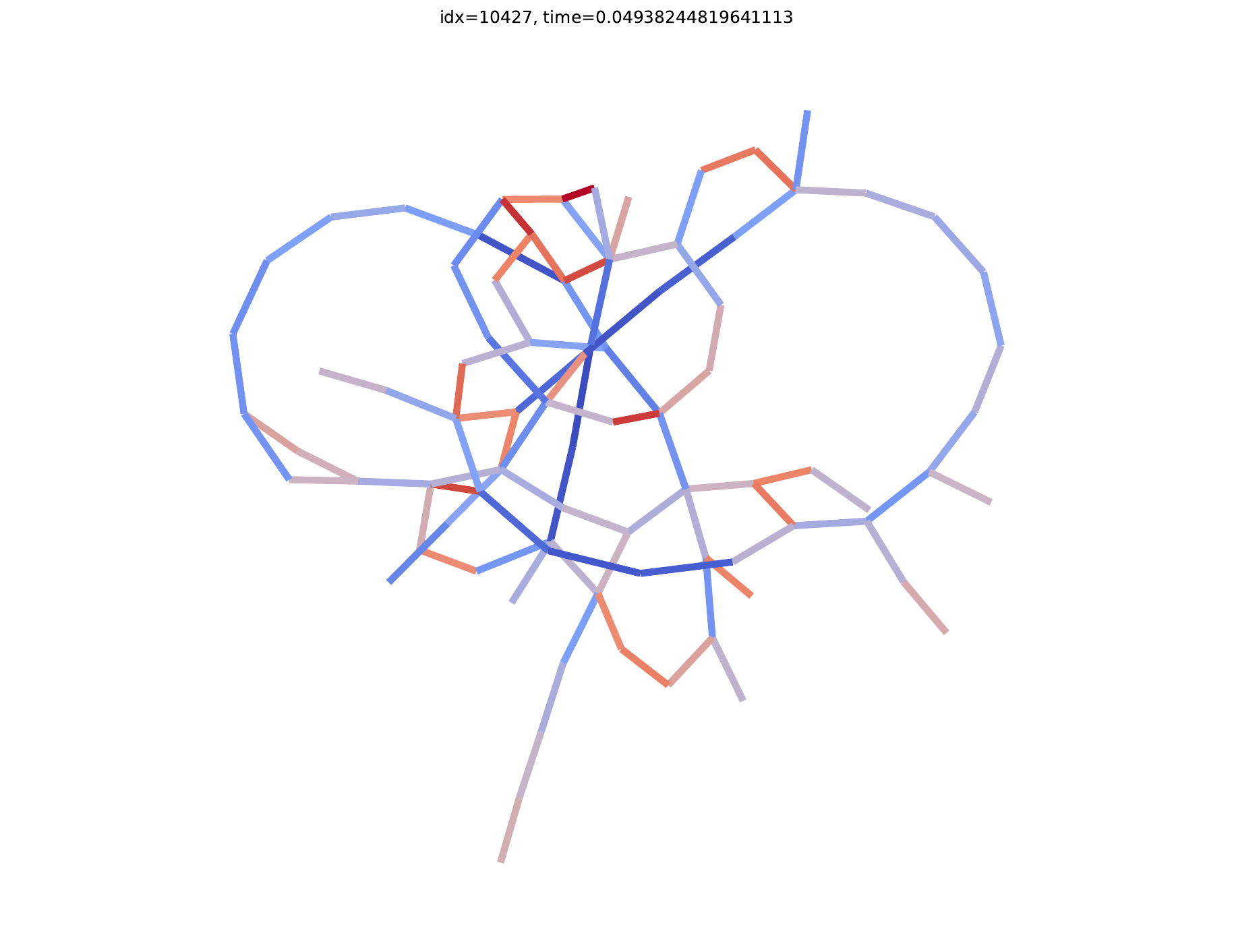} &
\imgcell{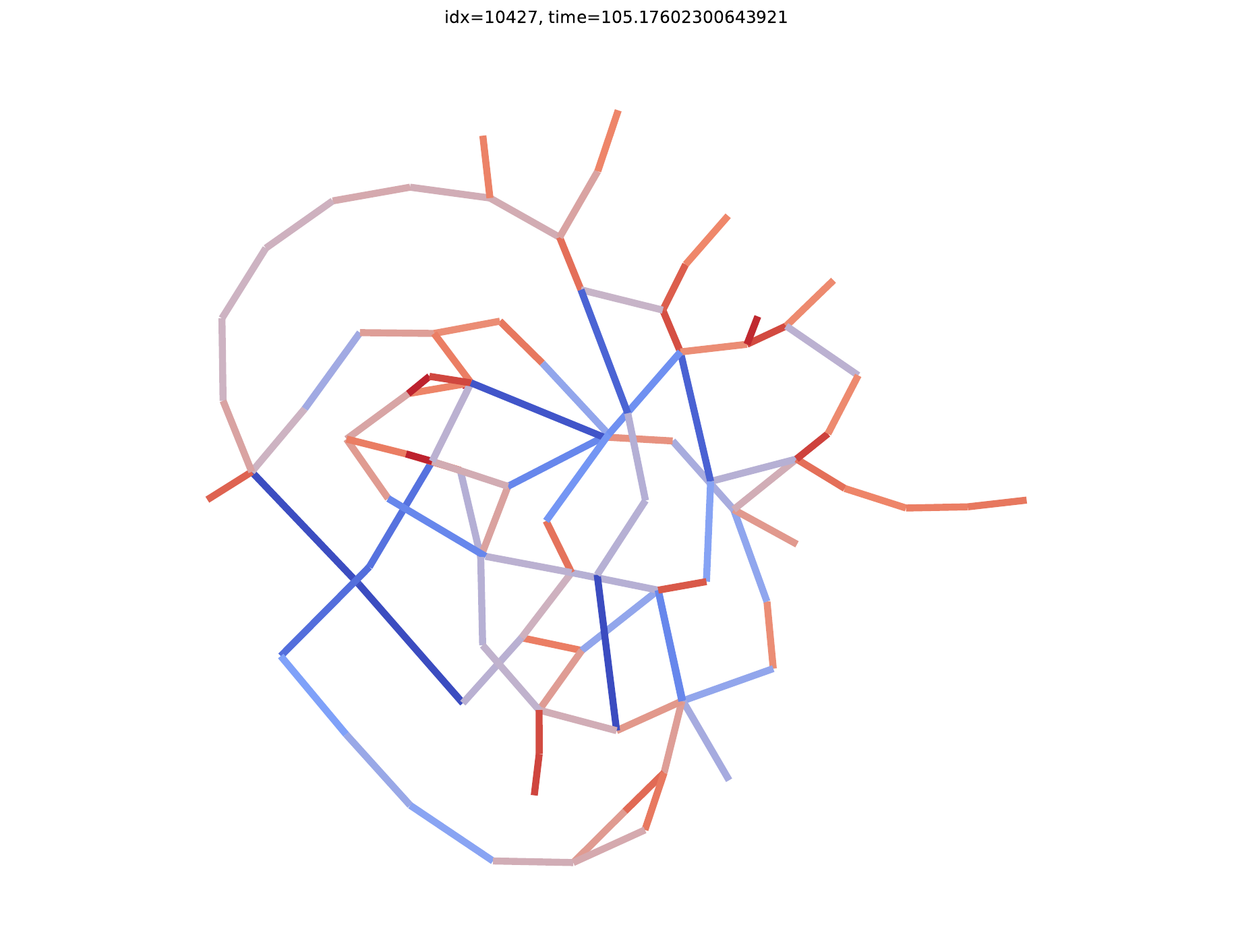} &
\imgcell{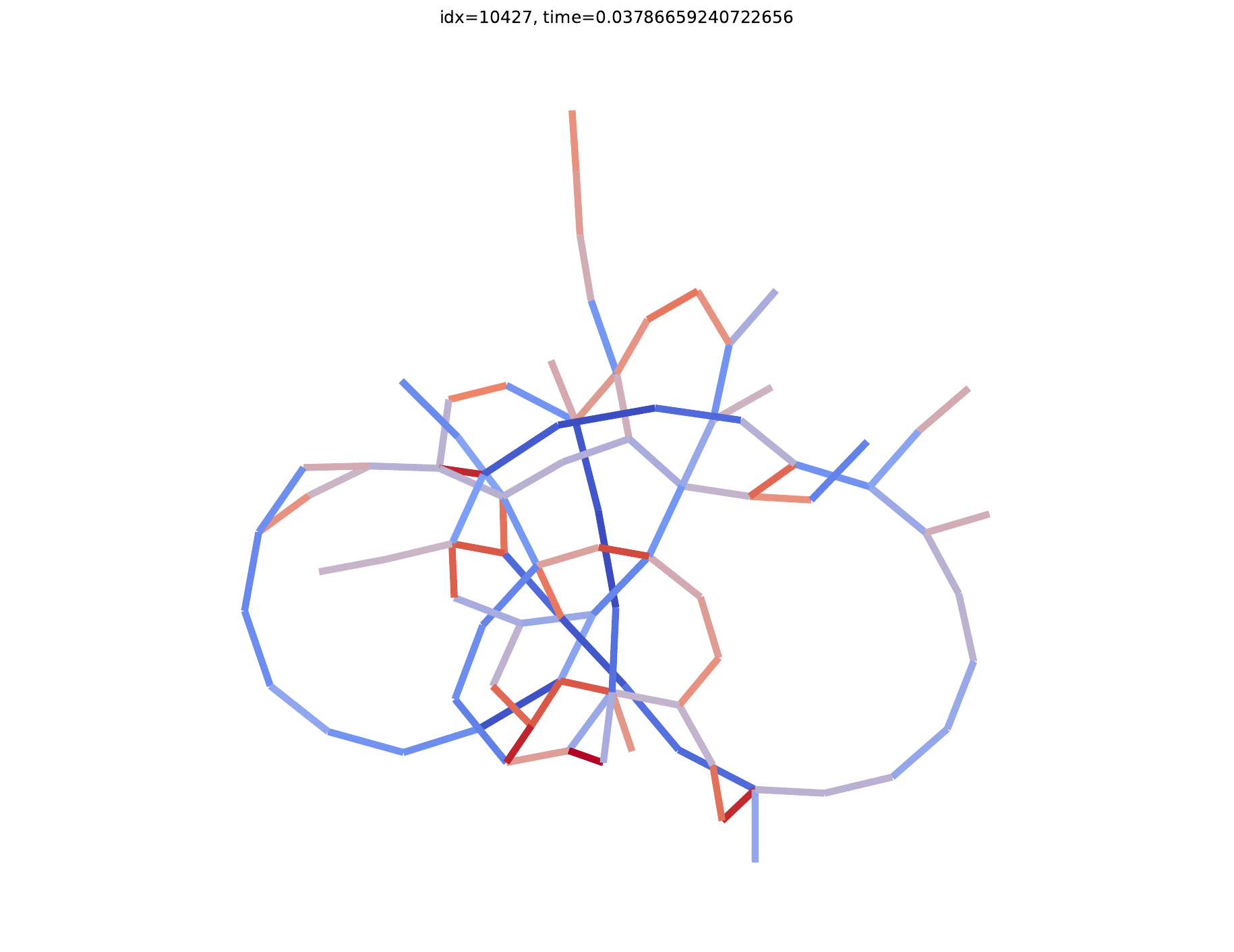} &
\imgcell{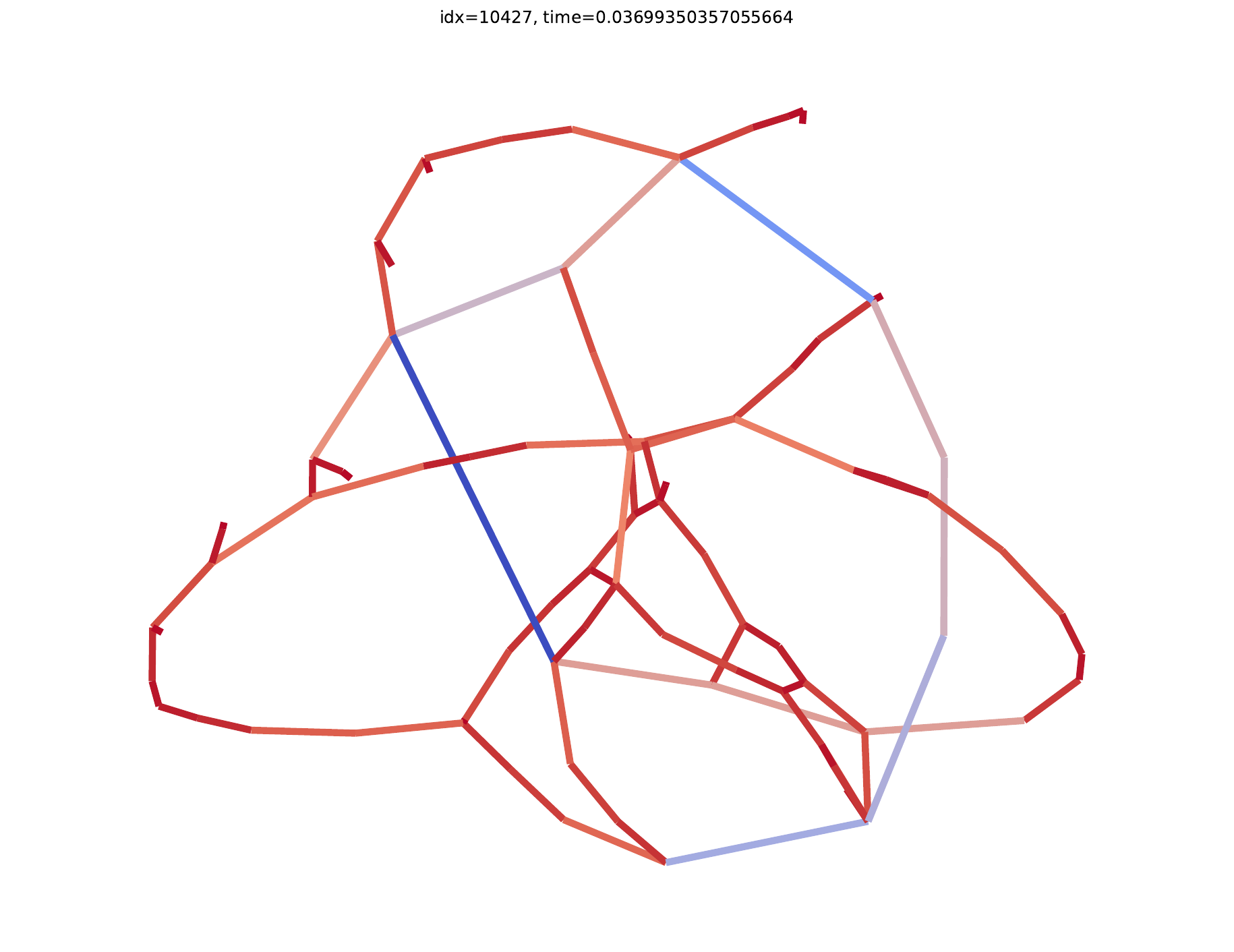} &
\imgcell{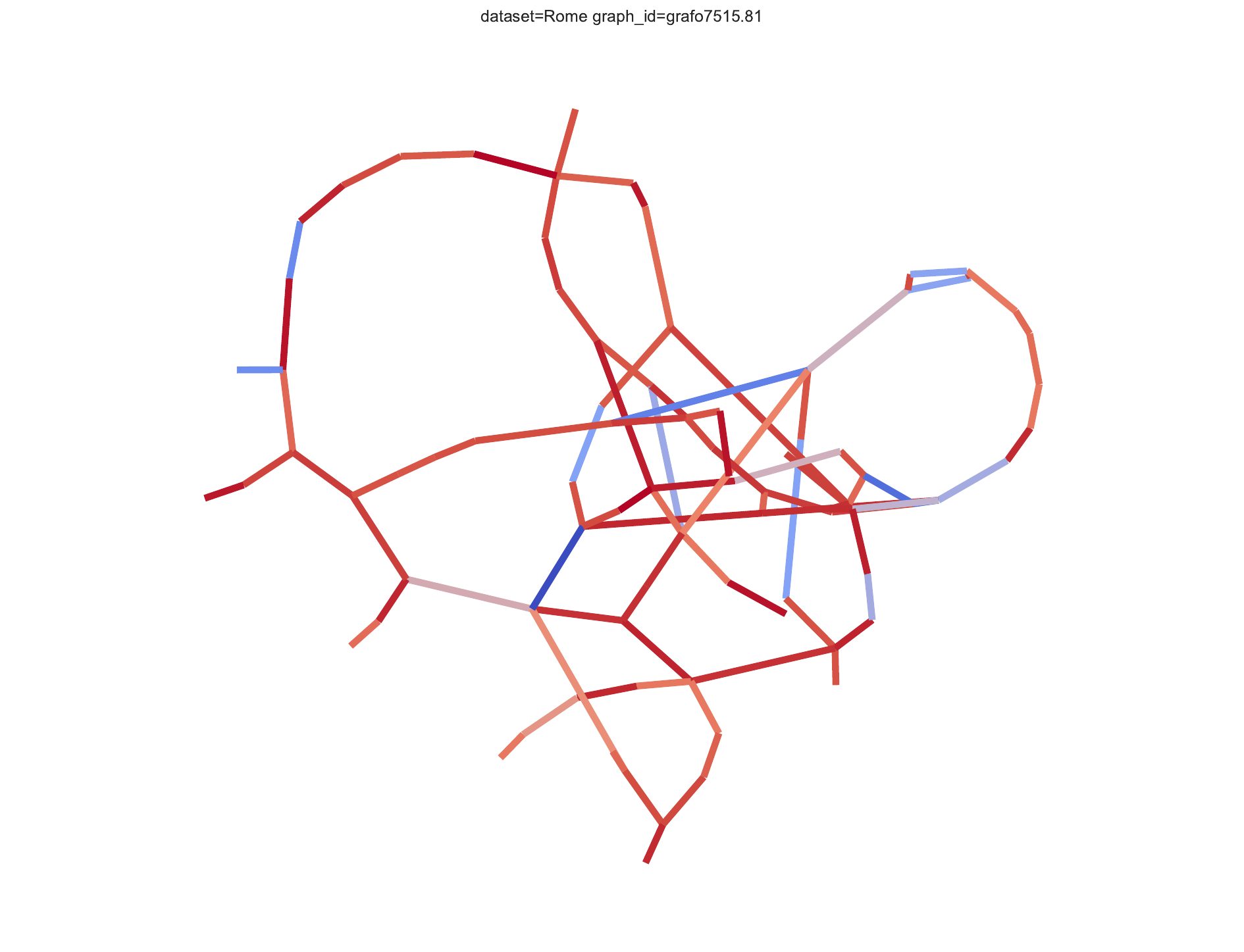} &
\imgcell{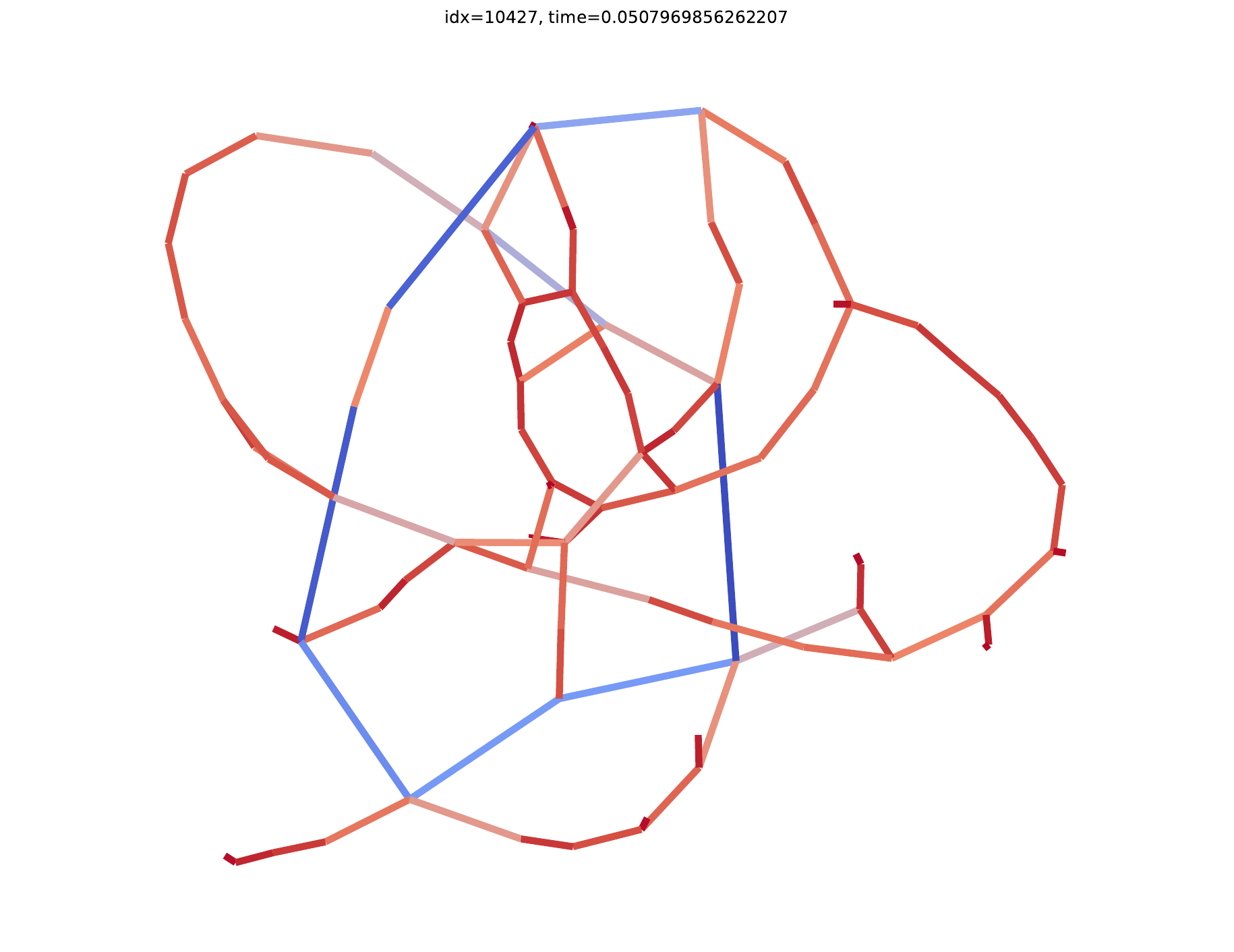} &
\imgcell{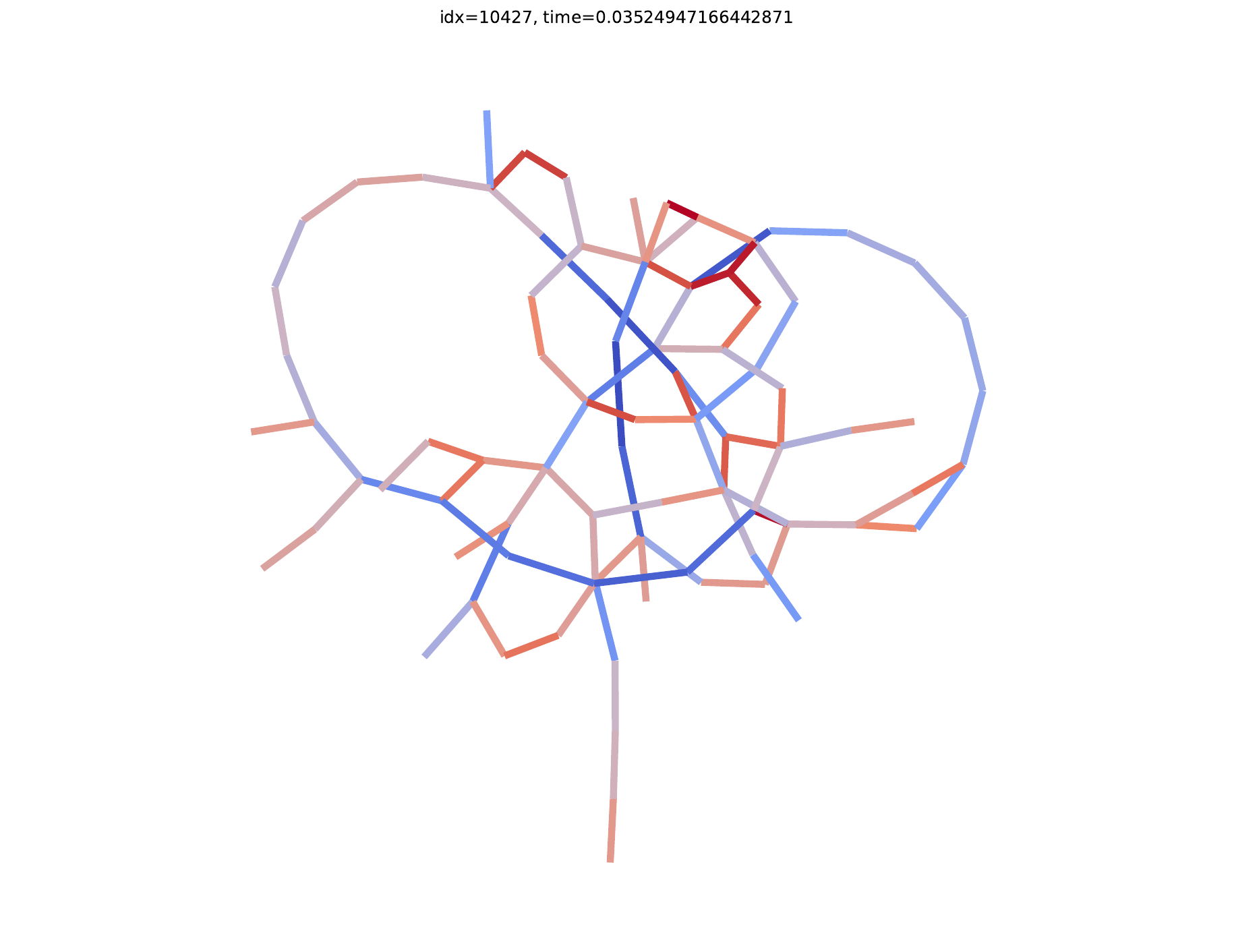} &
\imgcell{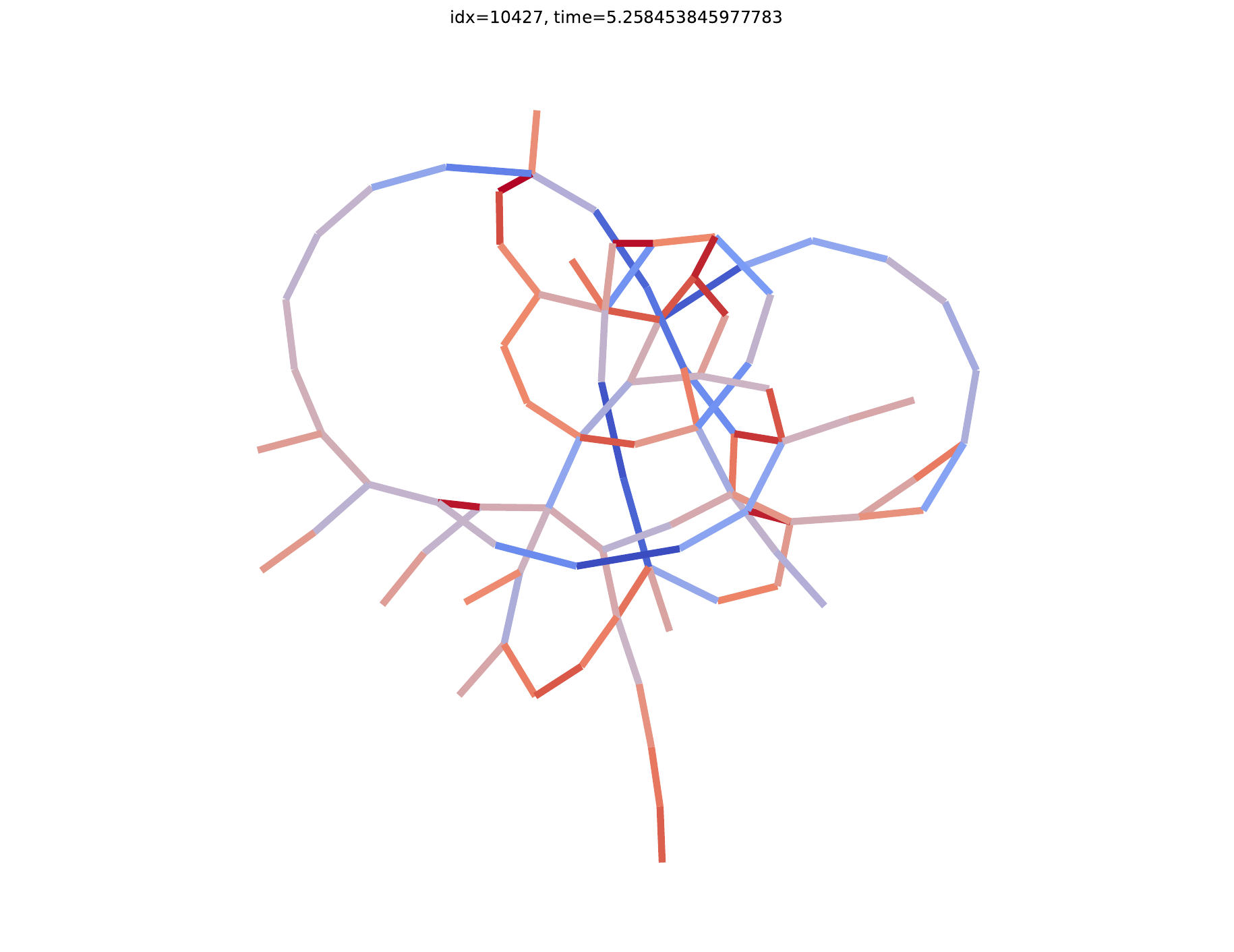} &
\imgcell{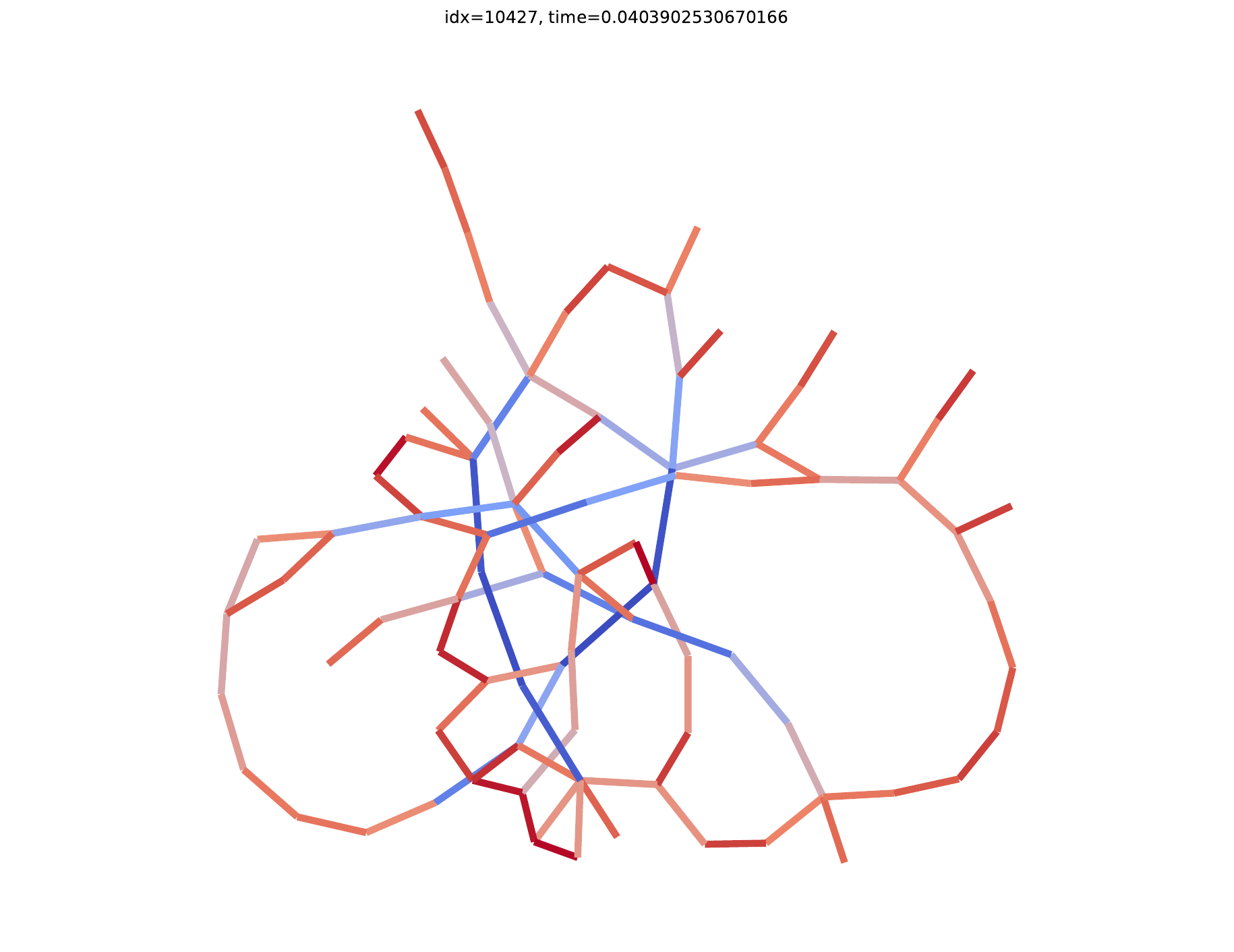} \\

&
t = 0.00s &
t = 1.70s &
t = 0.45s &
t = 0.05s &
t = 105.18s &
t = 0.04s &
t = 0.04s &
t = 0.04s &
t = 0.05s &
t = 0.04s &
t = 0.05s &
t = 0.04s \\

\makecell{\bfseries grafo10179.97\\N = 49\\M = 66} &
\imgcell{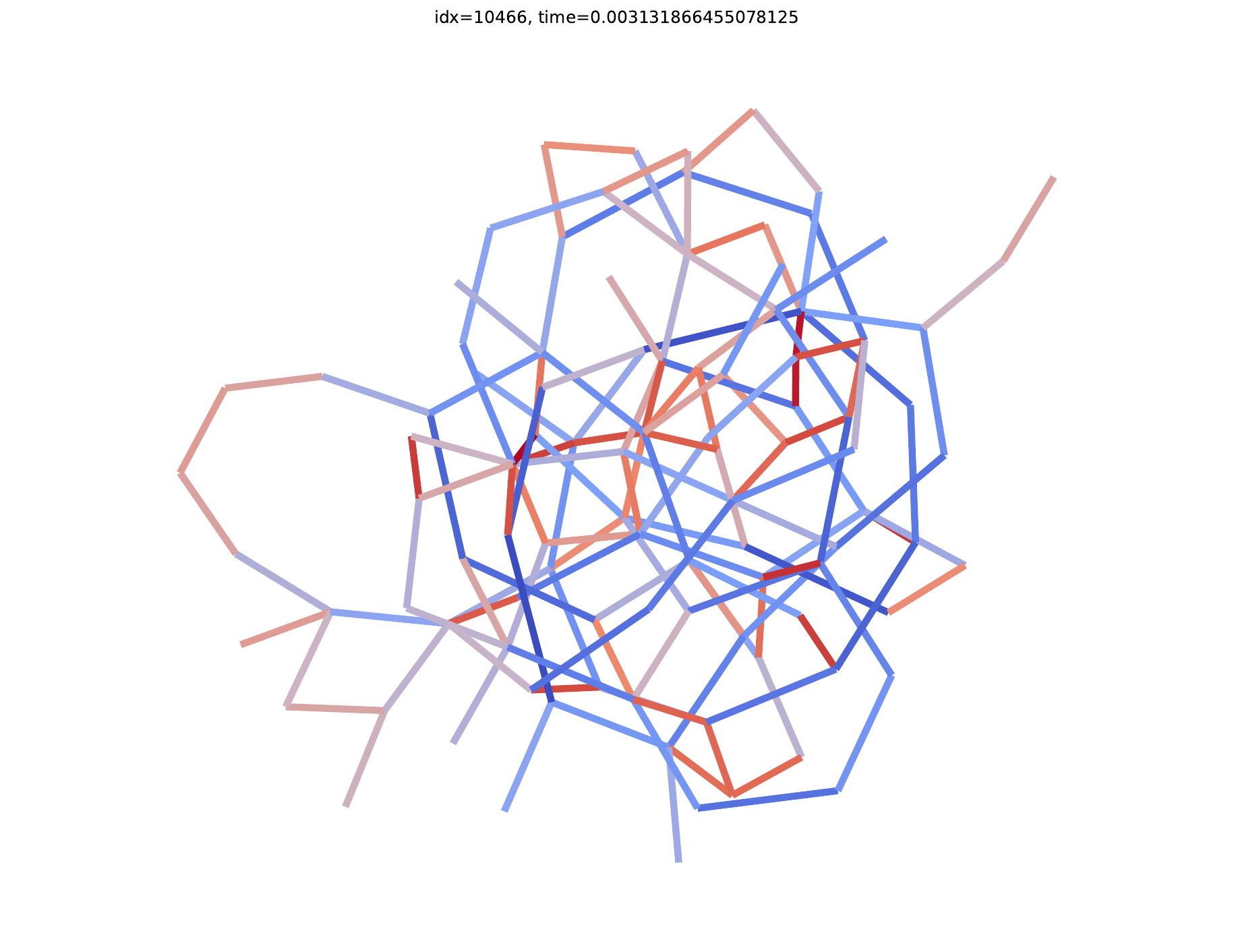} &
\imgcell{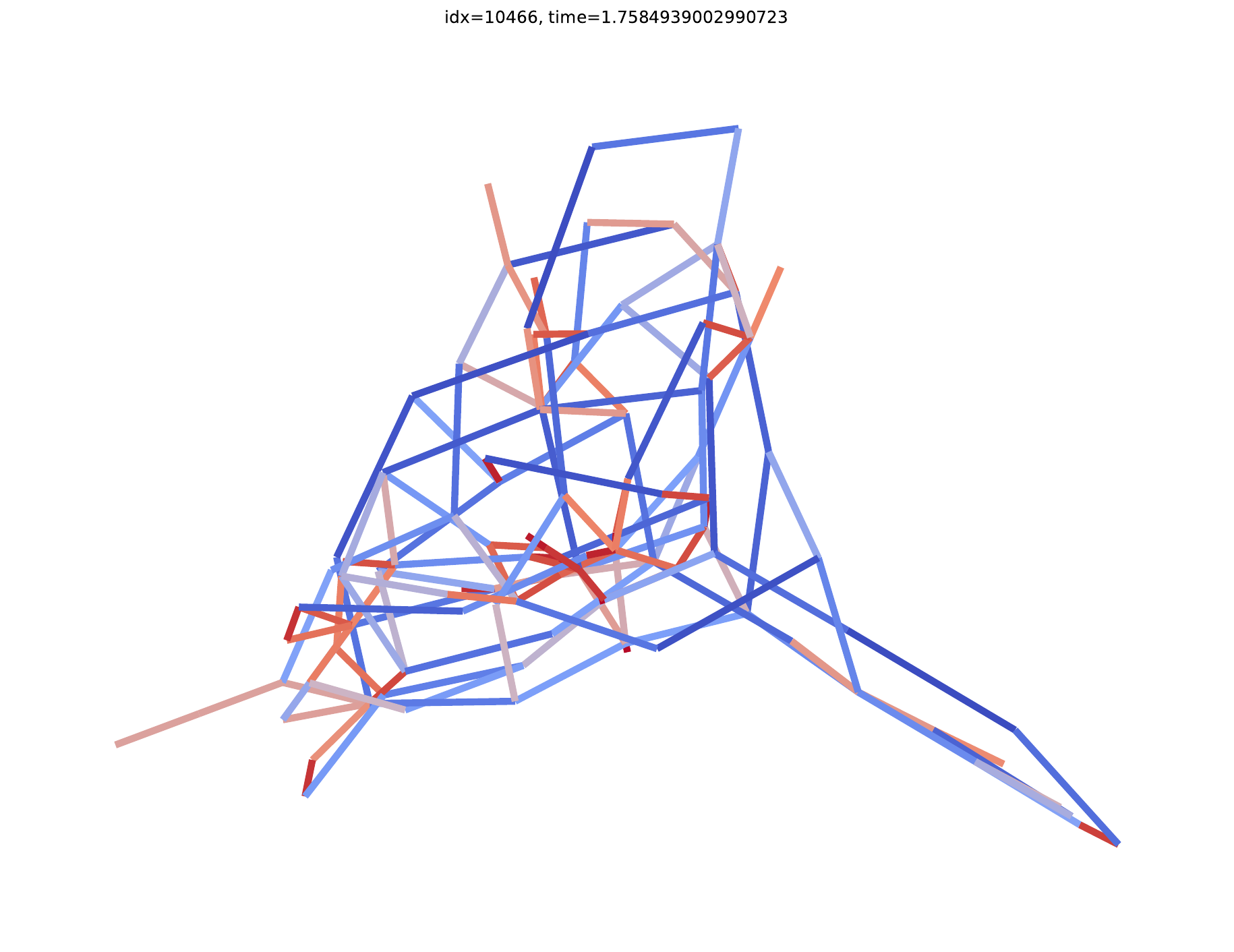} &
\imgcell{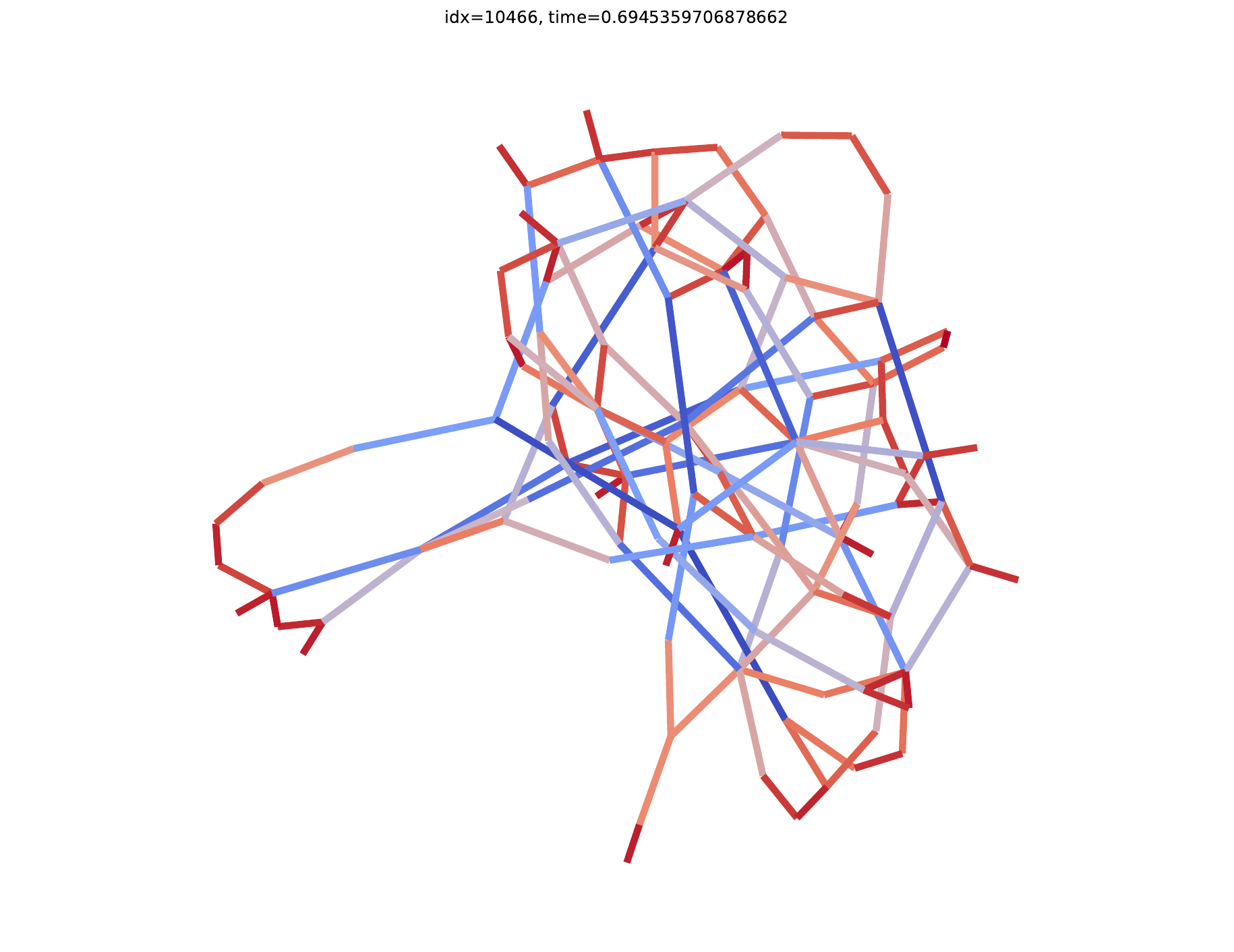} &
\imgcell{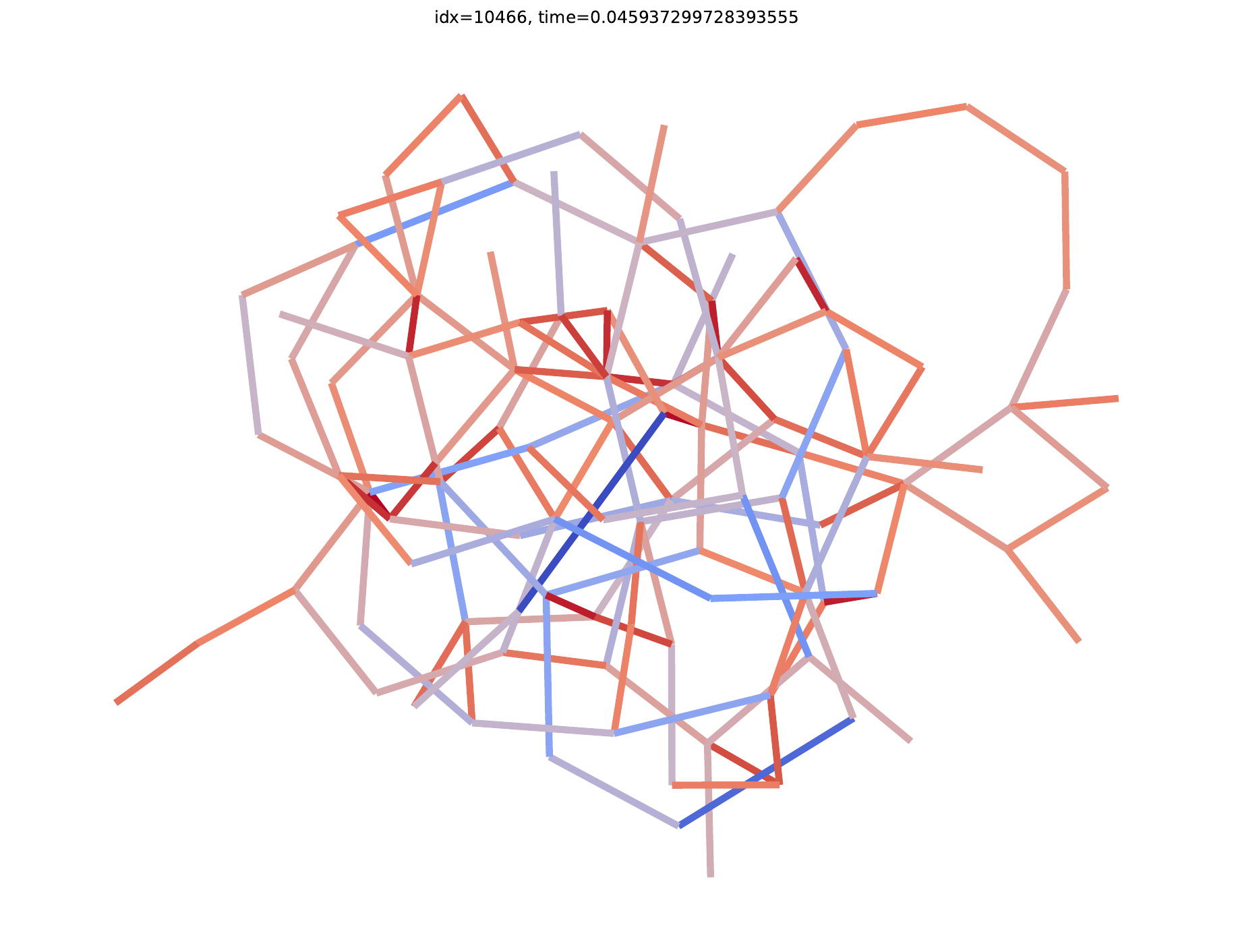} &
\imgcell{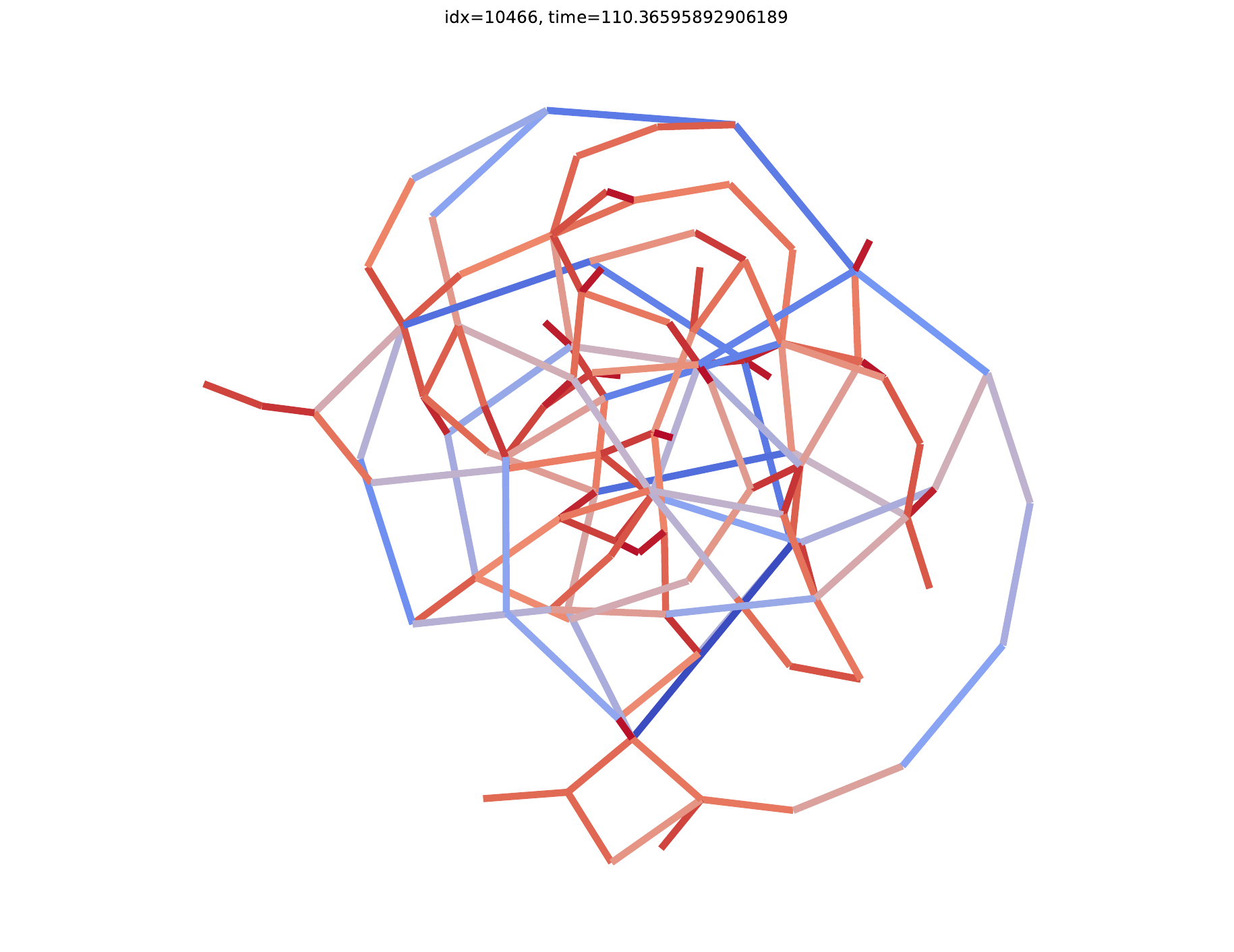} &
\imgcell{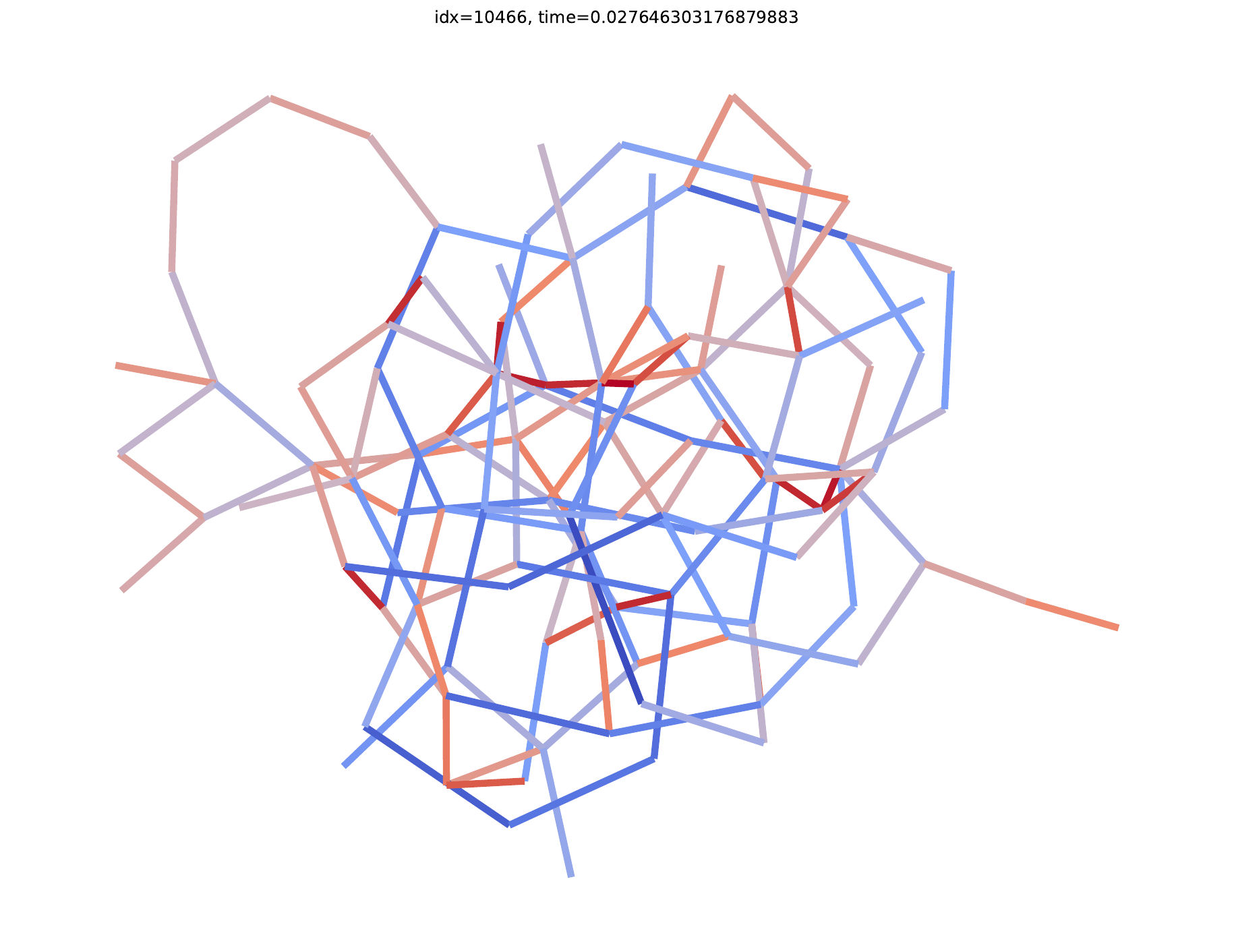} &
\imgcell{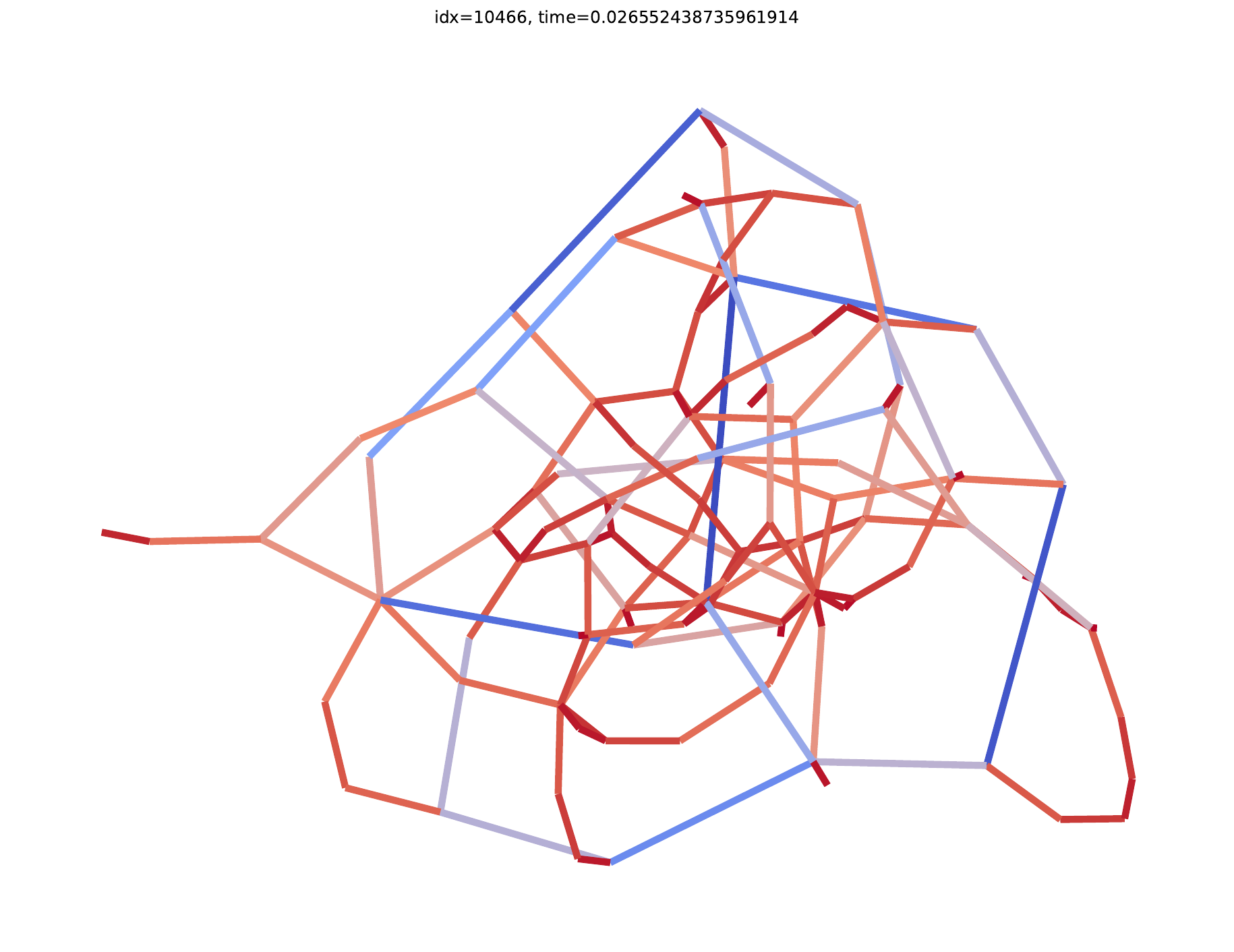} &
\imgcell{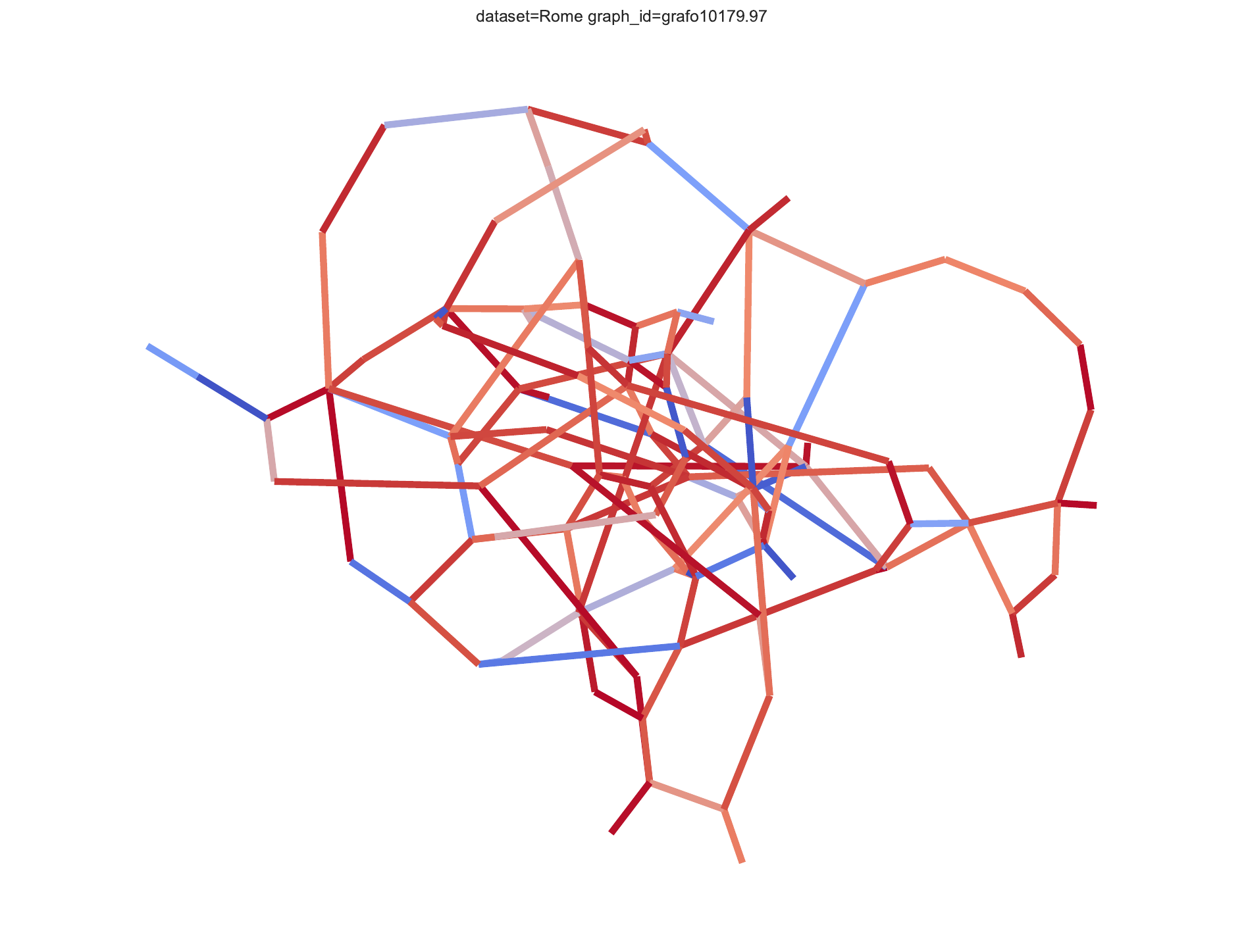} &
\imgcell{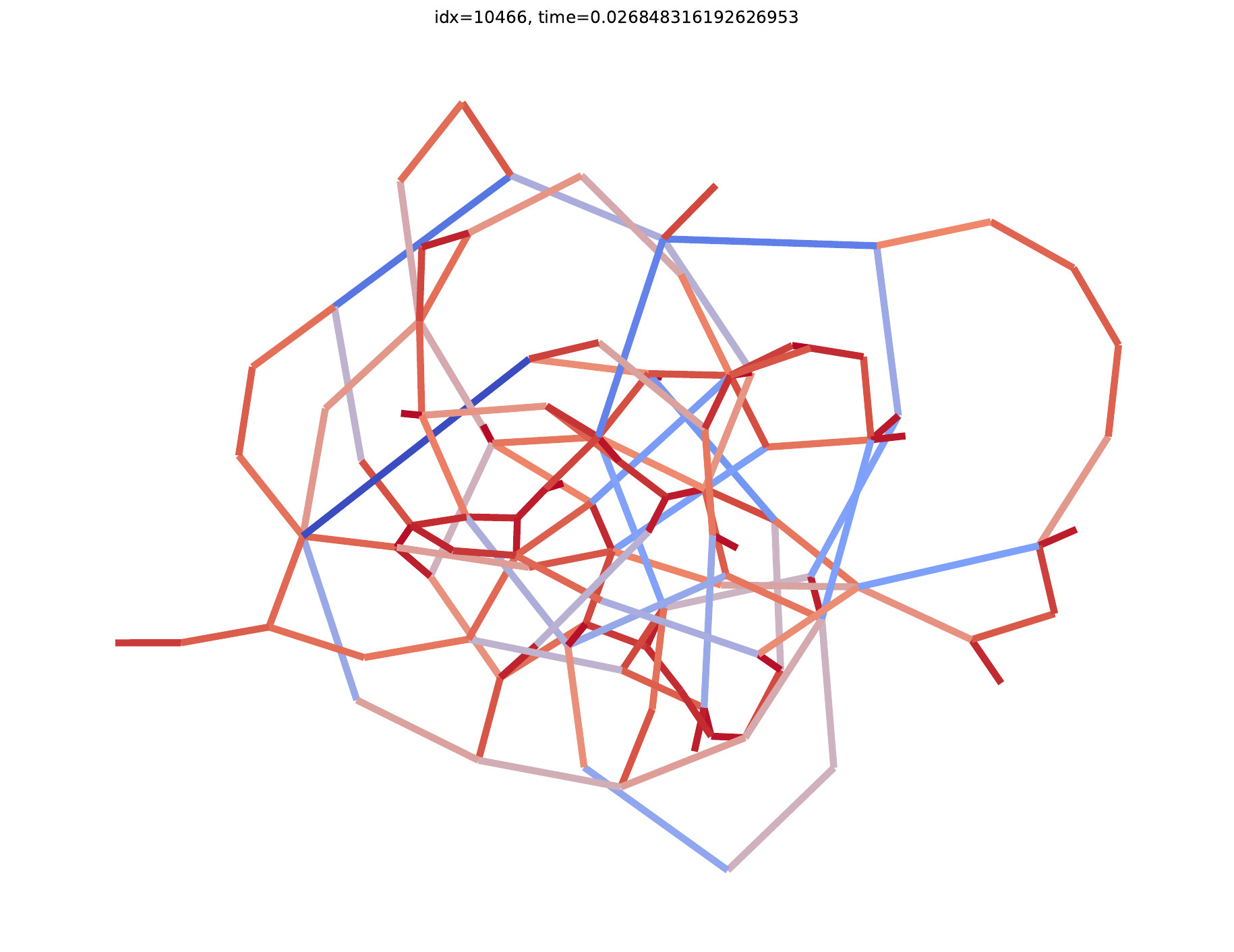} &
\imgcell{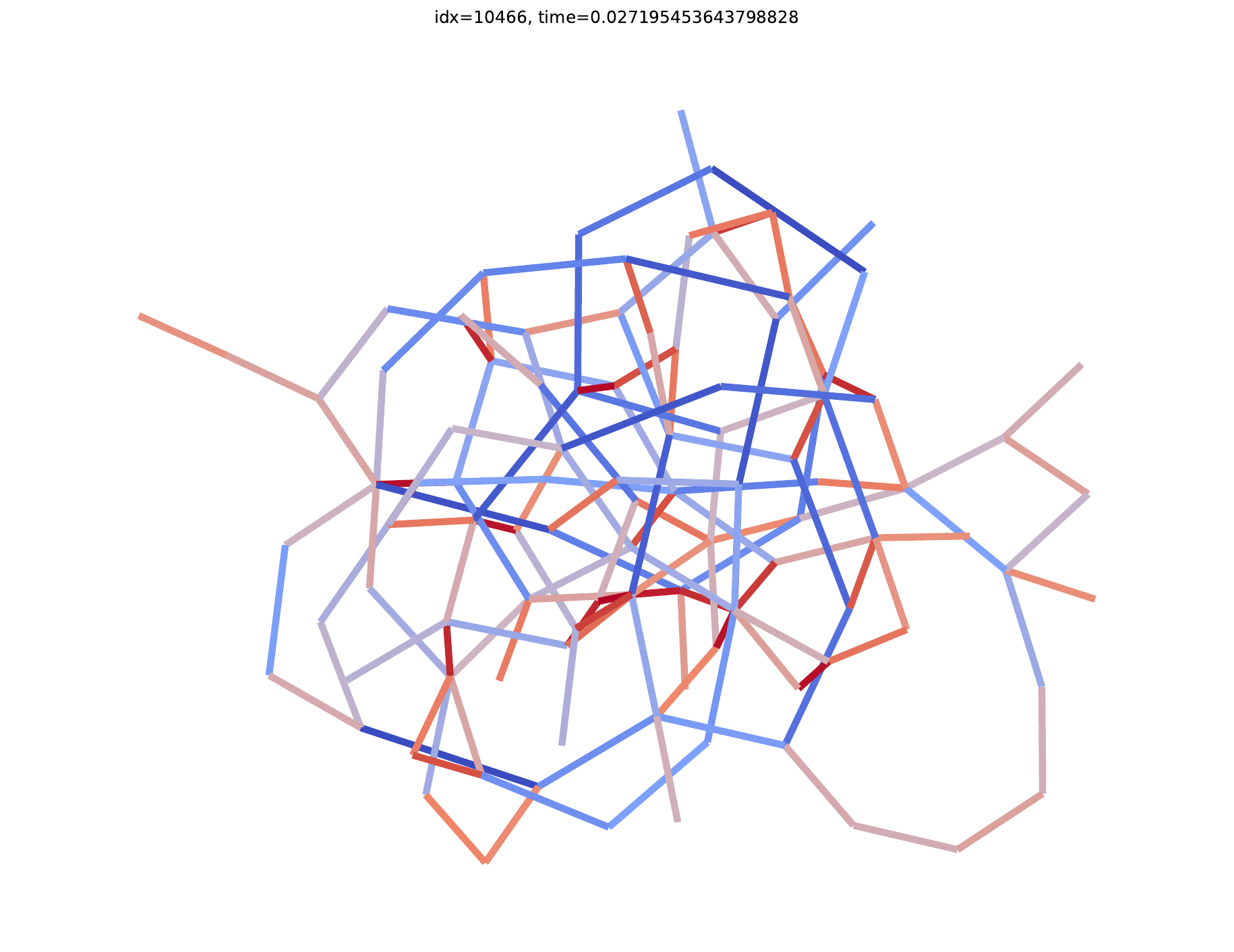} &
\imgcell{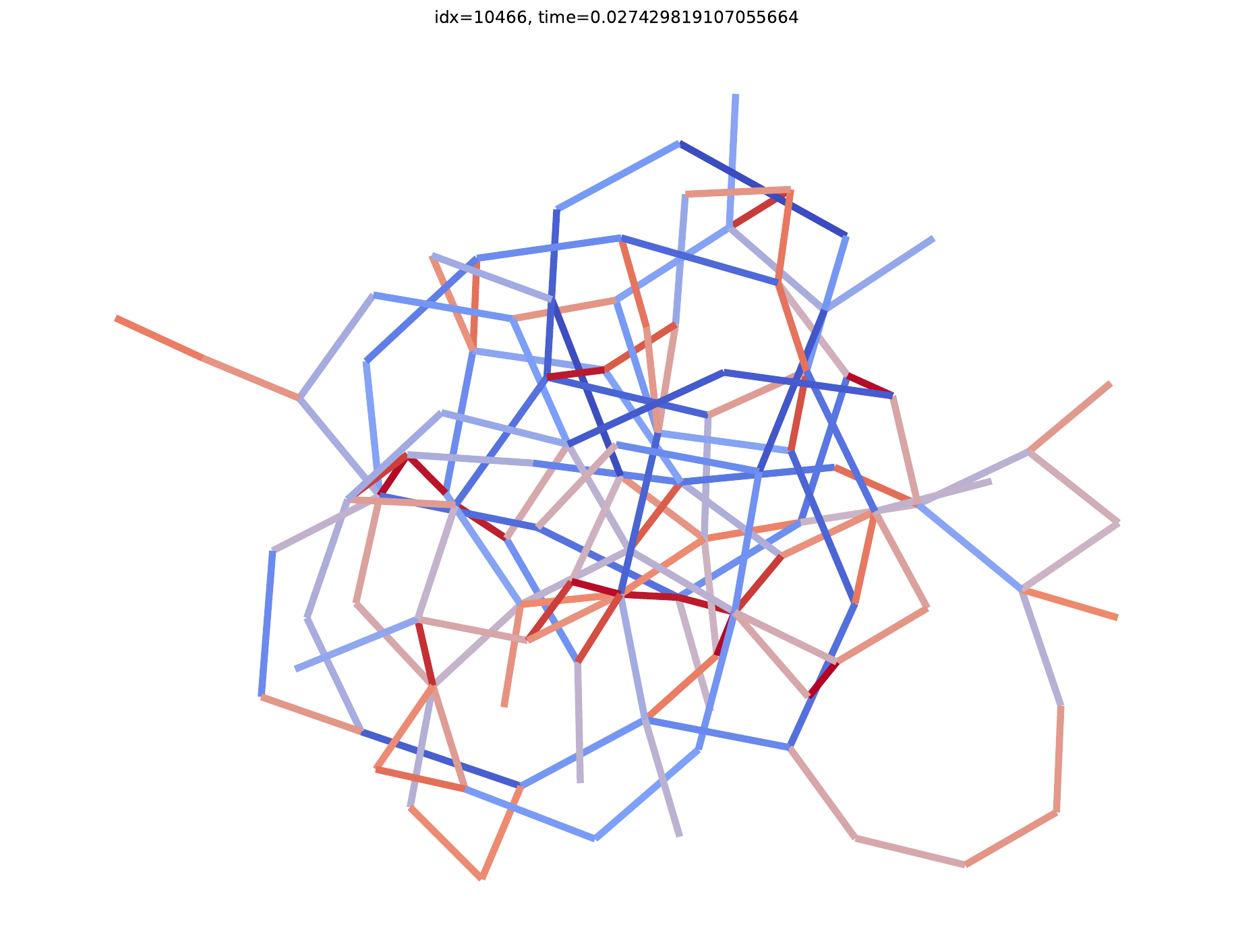} &
\imgcell{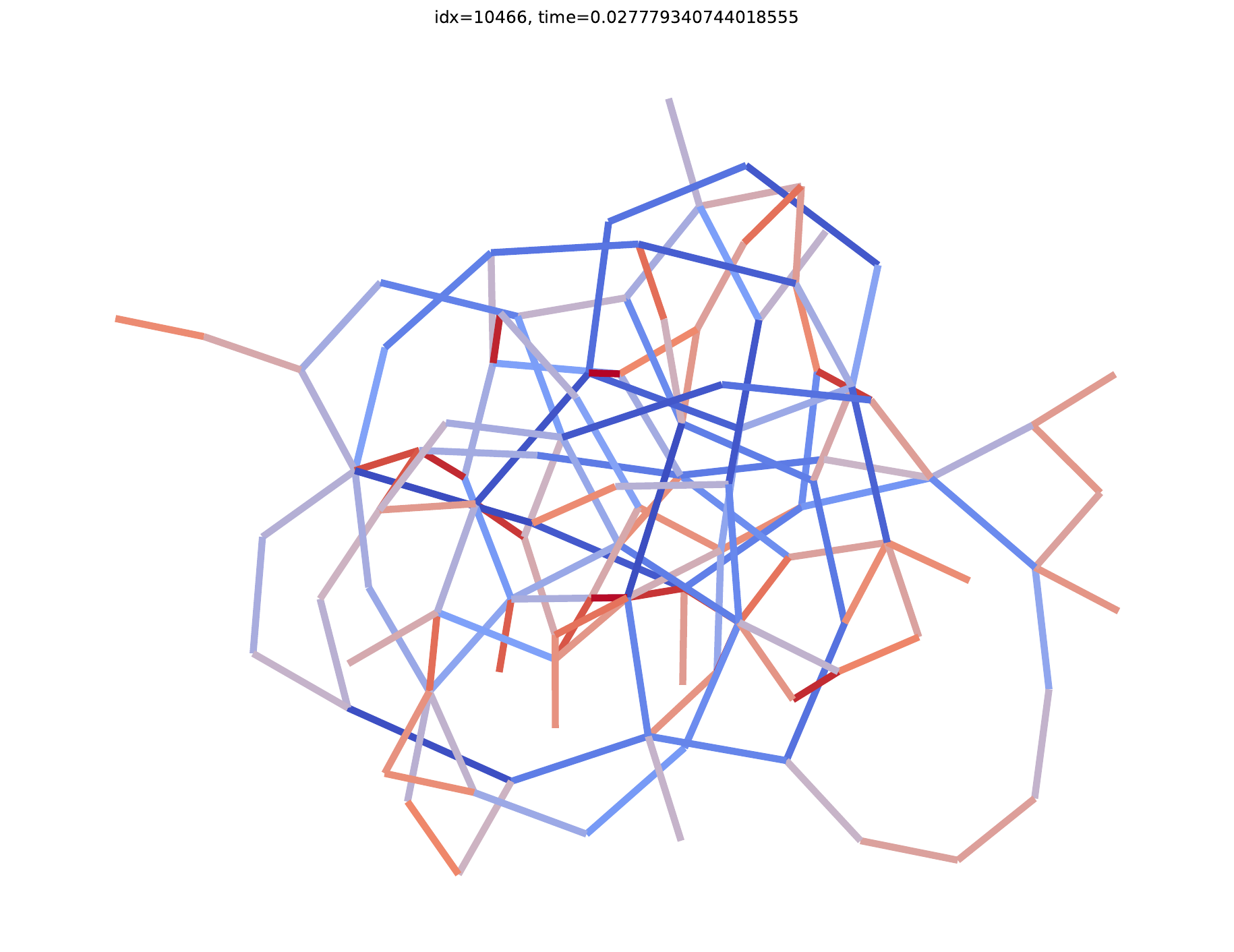} \\

&
t = 0.00s &
t = 1.76s &
t = 0.69s &
t = 0.05s &
t = 110.37s &
t = 0.03s &
t = 0.03s &
t = 0.04s &
t = 0.03s &
t = 0.03s &
t = 0.03s &
t = 0.03s \\

\makecell{\bfseries grafo5807.32\\N = 95\\M = 121} &
\imgcell{figures/rome_graphs/10496_sgd2.pdf} &
\imgcell{figures/rome_graphs/10496_pmds.pdf} &
\imgcell{figures/rome_graphs/10496_fa2.pdf} &
\imgcell{figures/rome_graphs/10496_deepgd.pdf} &
\imgcell{figures/rome_graphs/10496_gd2_stress_xing.pdf} &
\imgcell{figures/rome_graphs/10496_smartgd_stress.pdf} &
\imgcell{figures/rome_graphs/10496_smartgd_xing.pdf} &
\imgcell{figures/rome_graphs/10496_smartgd_xing_nsc.pdf} &
\imgcell{figures/rome_graphs/10496_smartgd_xangle.pdf} &
\imgcell{figures/rome_graphs/10496_smartgd_stress_xing.pdf} &
\imgcell{figures/rome_graphs/10496_smartgd_stress_xangle.pdf} &
\imgcell{figures/rome_graphs/10496_smartgd_combined.pdf} \\

&
t = 0.00s &
t = 0.32s &
t = 0.08s &
t = 0.05s &
t = 104.88s &
t = 0.04s &
t = 0.03s &
t = 0.04s &
t = 0.05s &
t = 0.03s &
t = 0.03s &
t = 0.03s \\

\makecell{\bfseries grafo8688.89\\N = 48\\M = 69} &
\imgcell{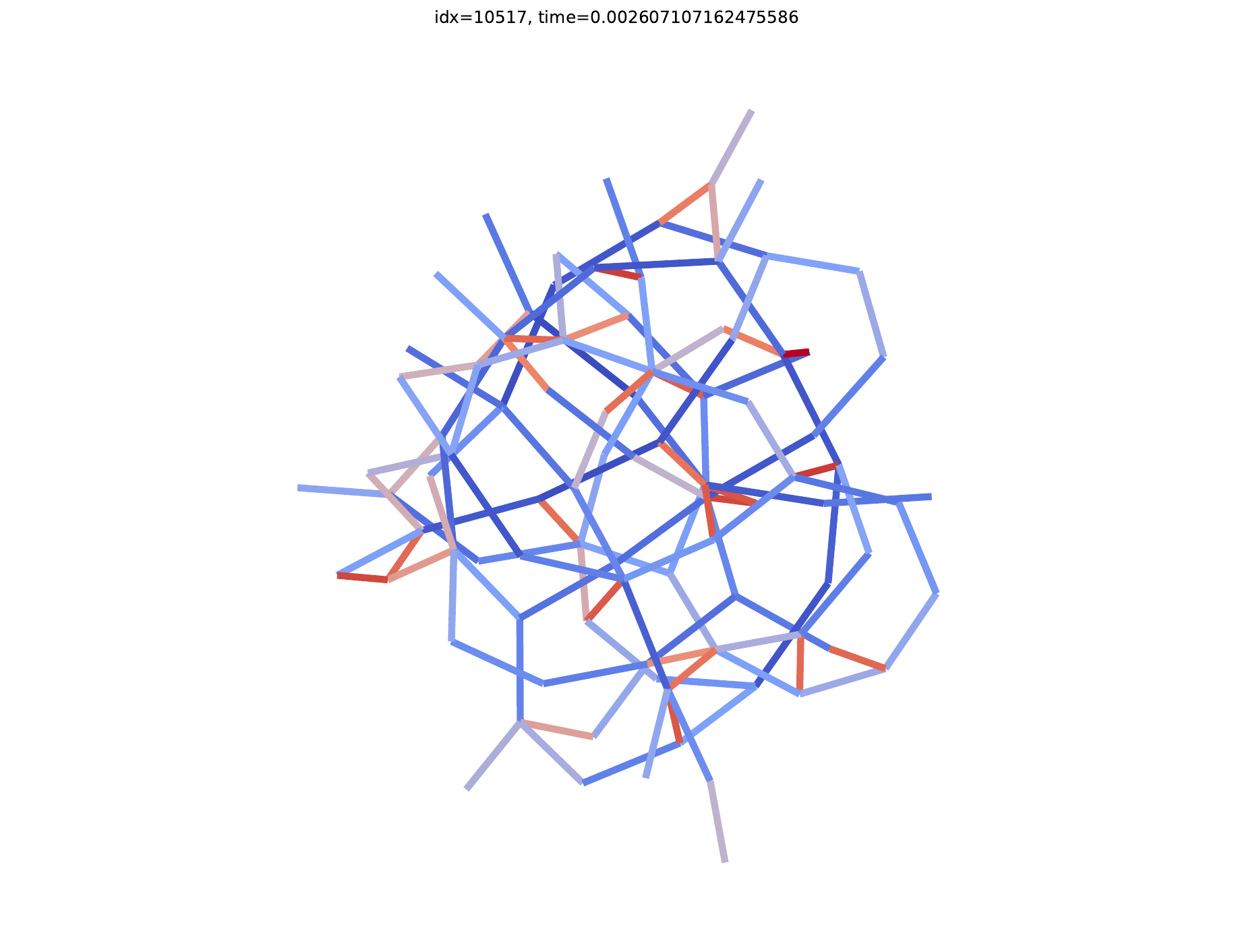} &
\imgcell{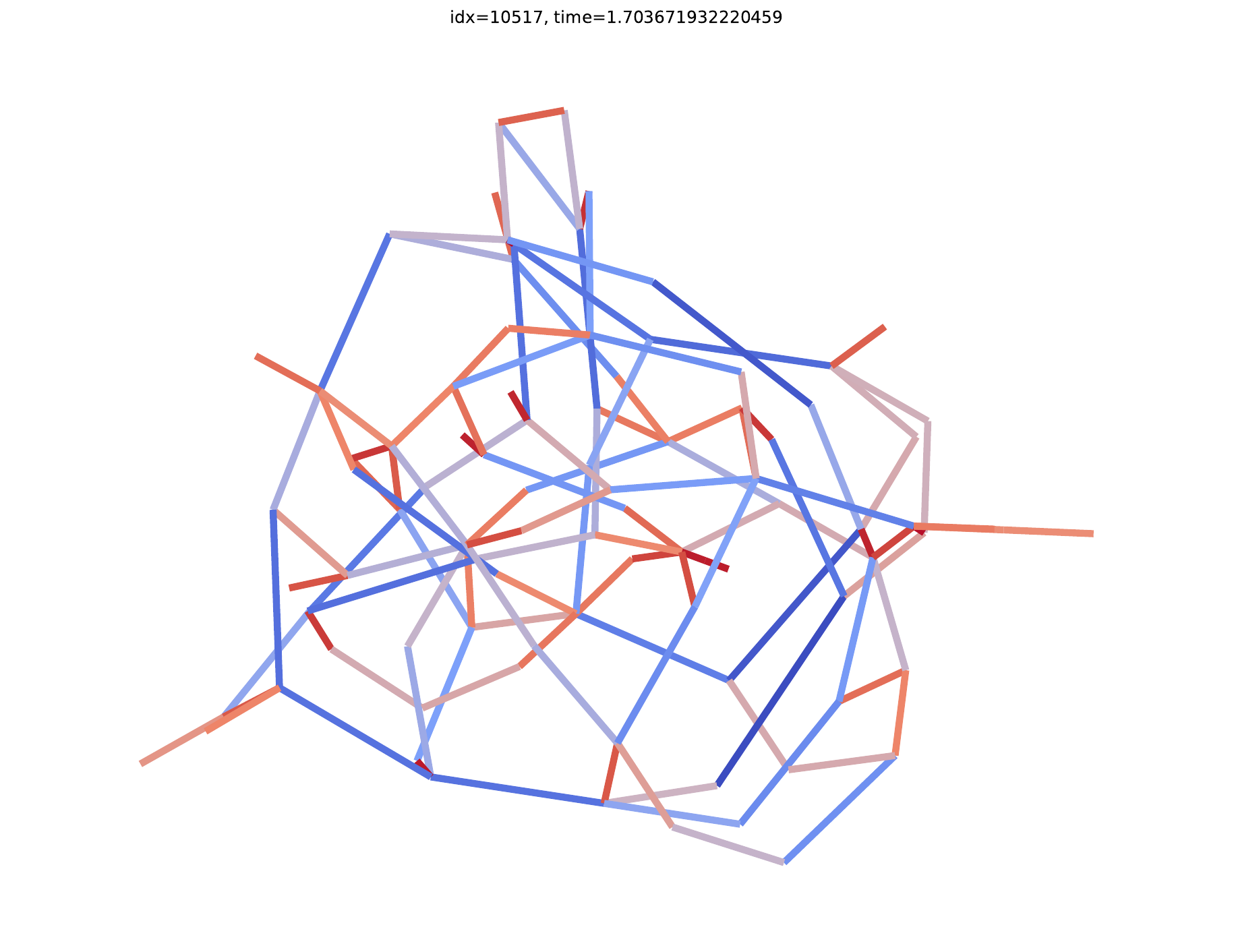} &
\imgcell{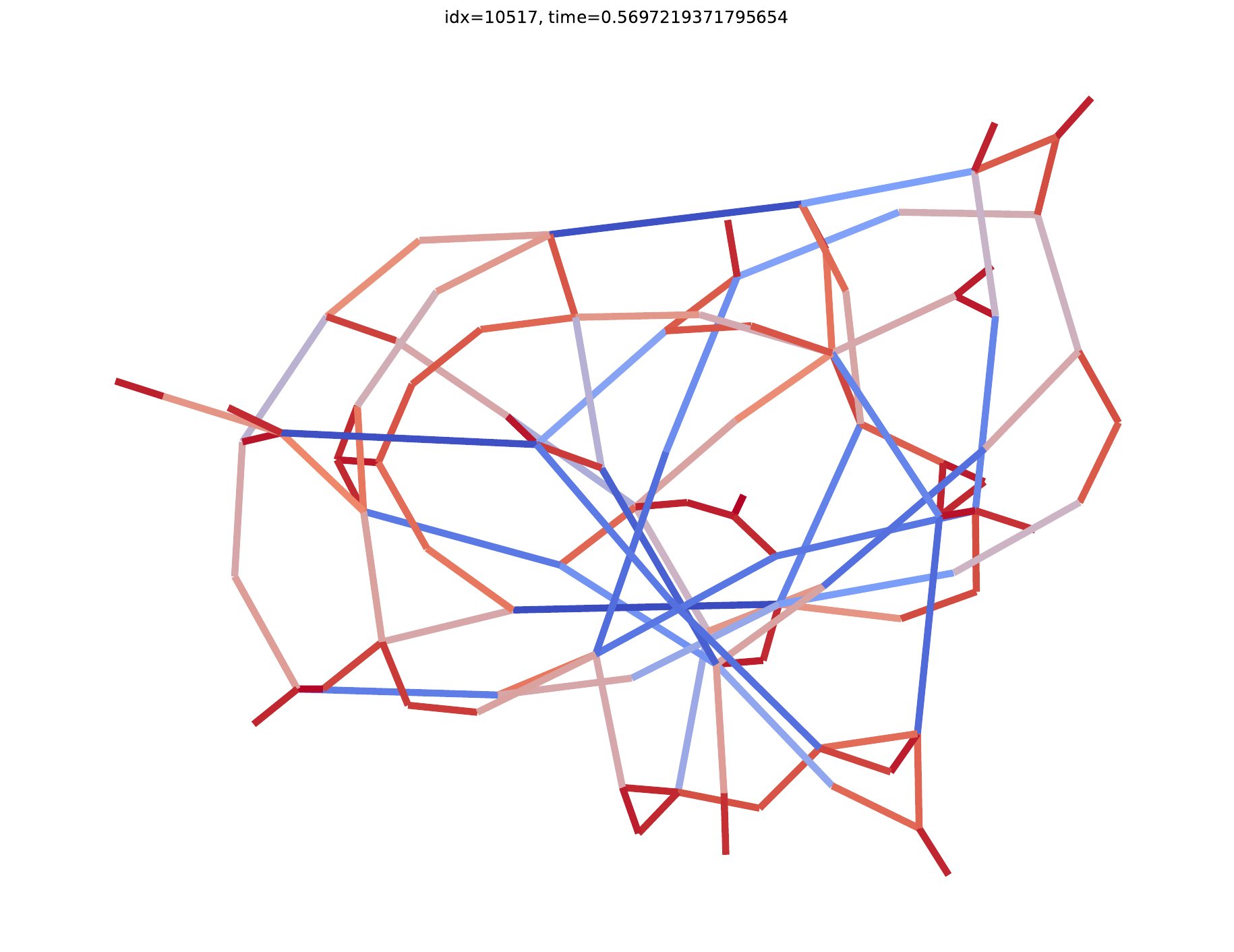} &
\imgcell{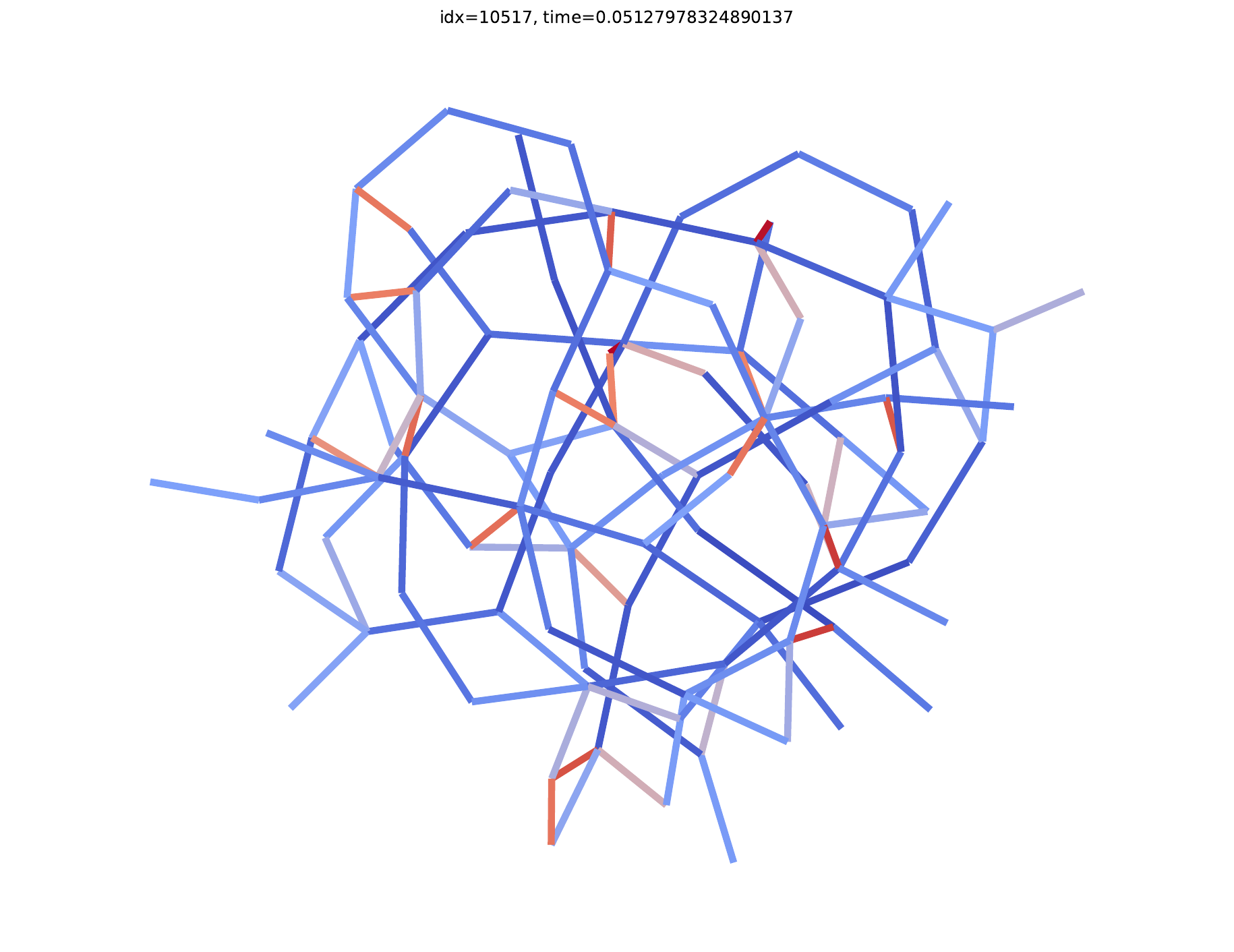} &
\imgcell{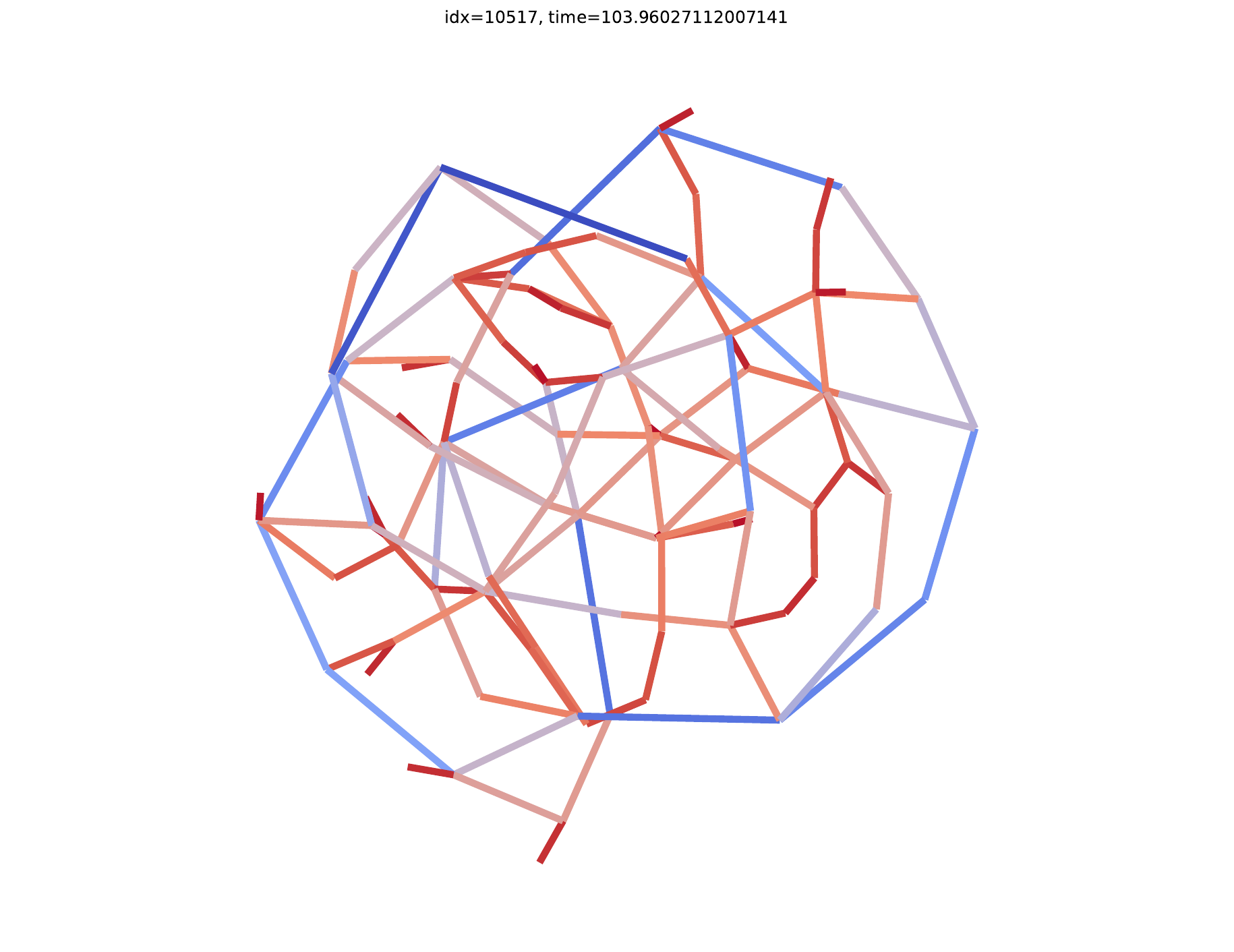} &
\imgcell{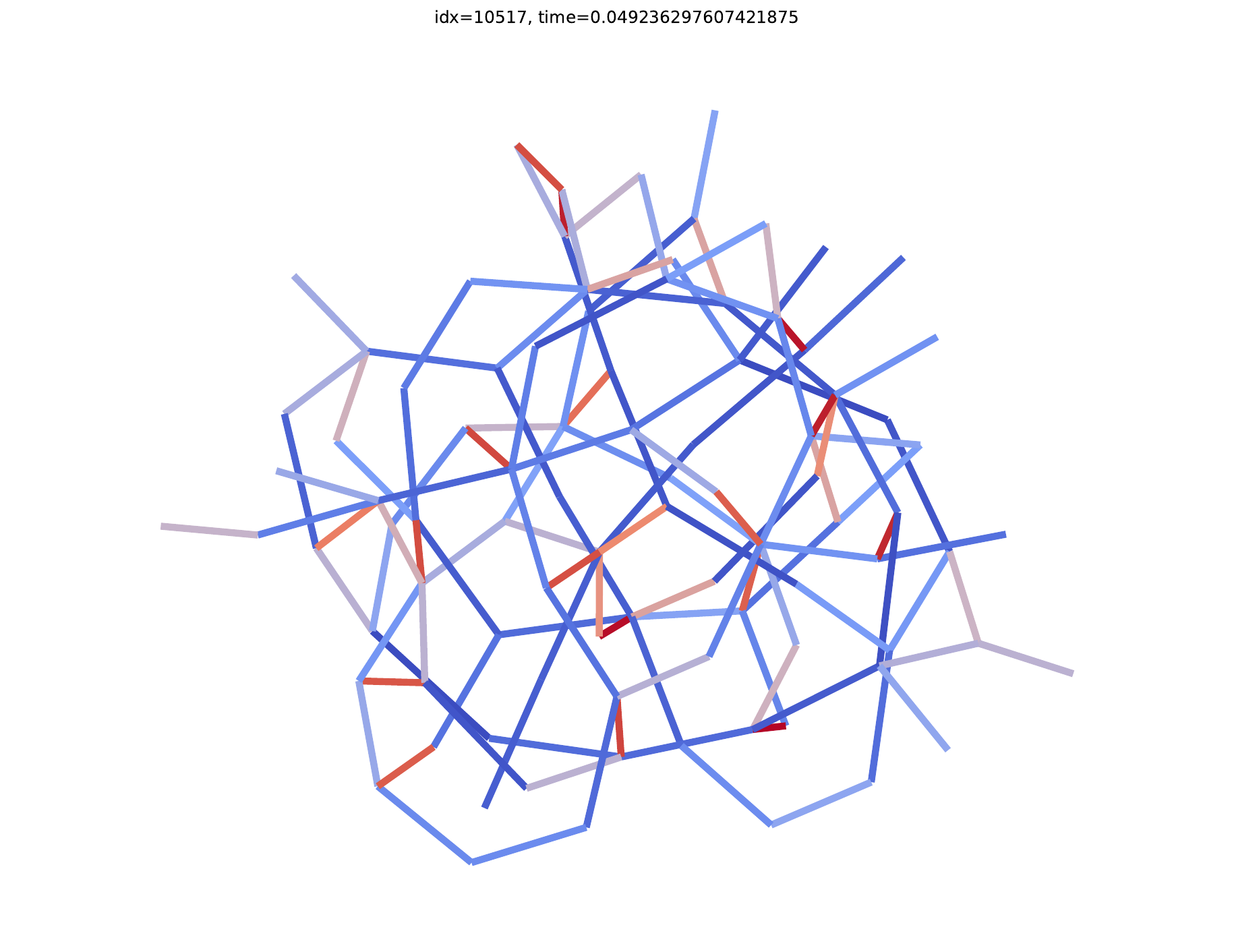} &
\imgcell{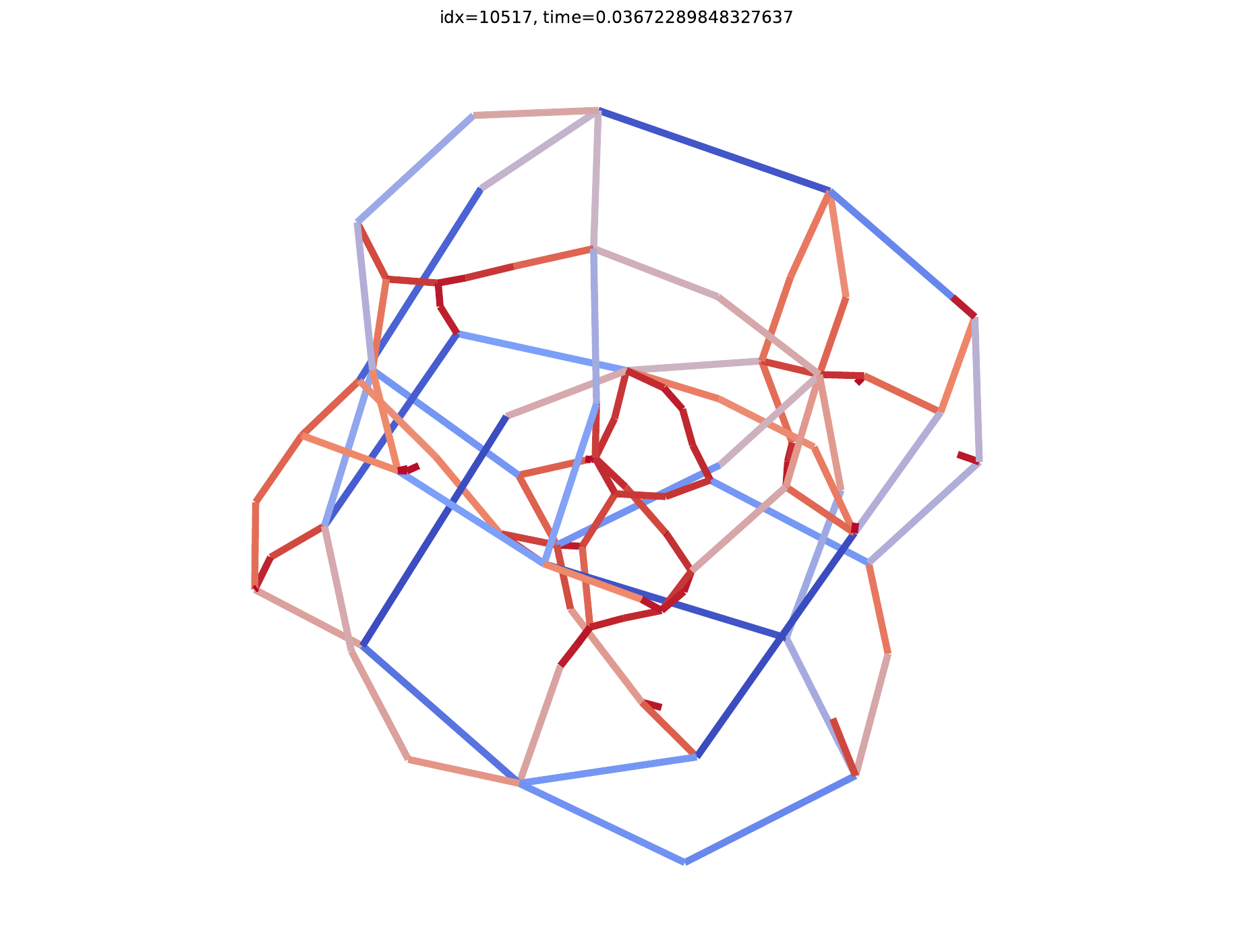} &
\imgcell{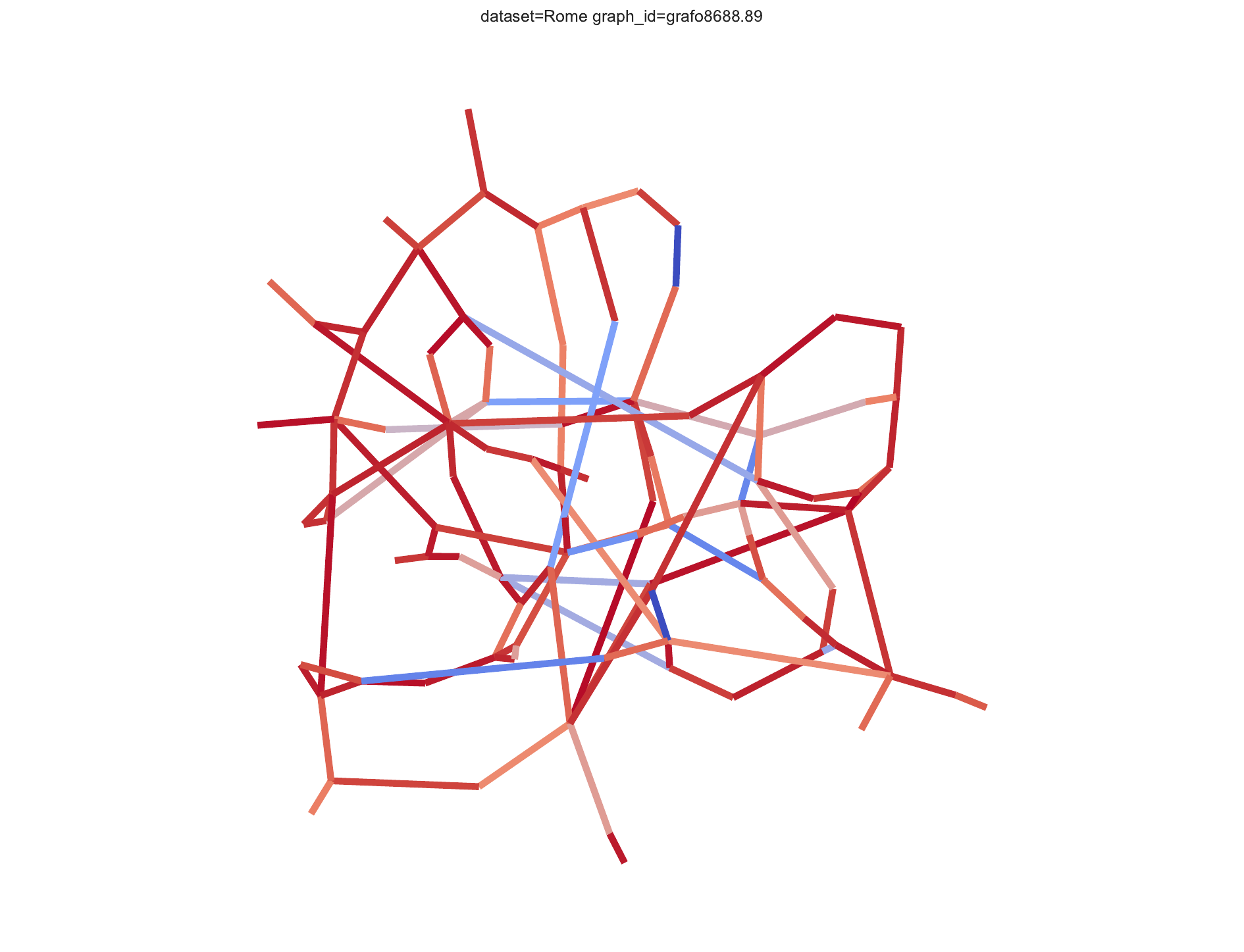} &
\imgcell{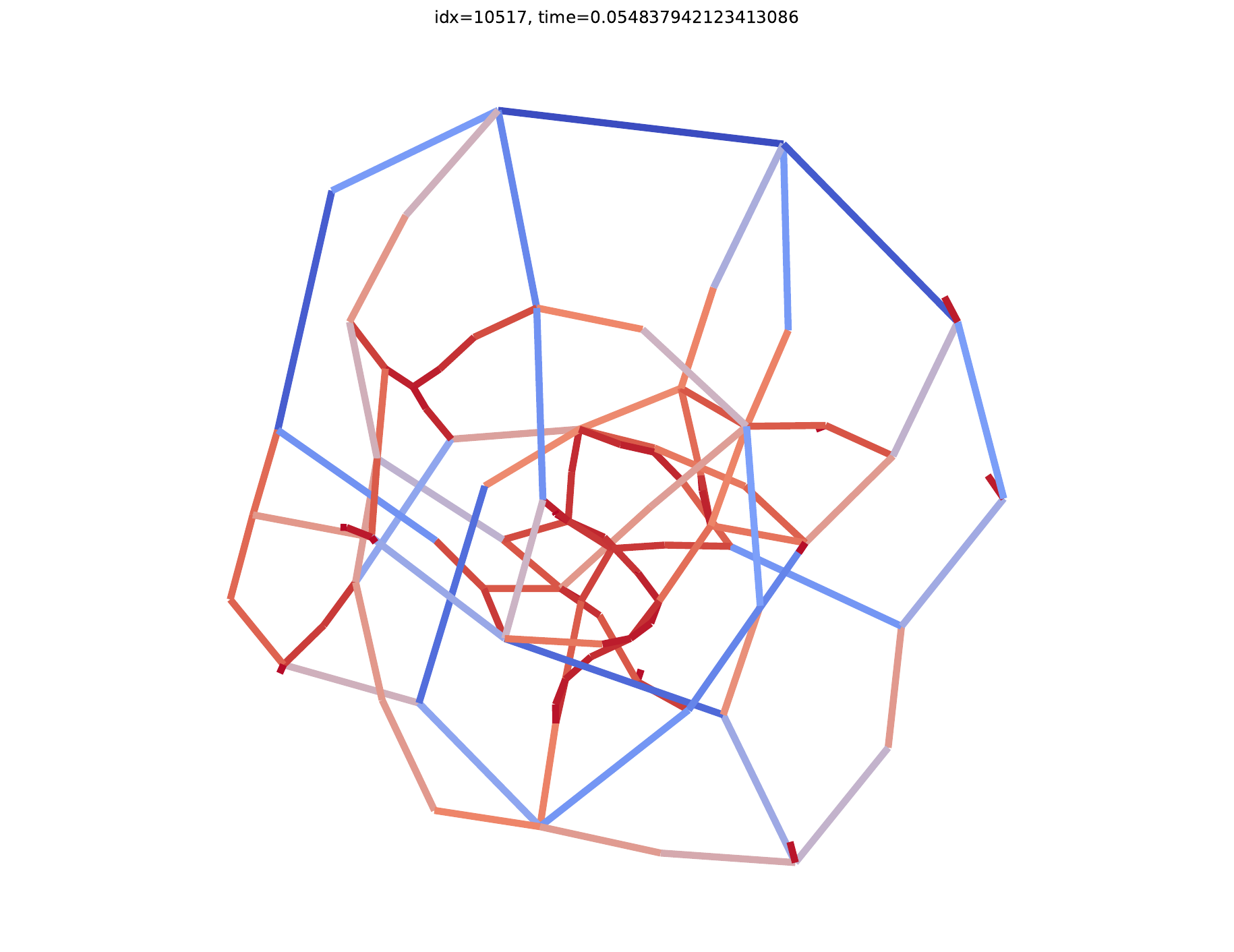} &
\imgcell{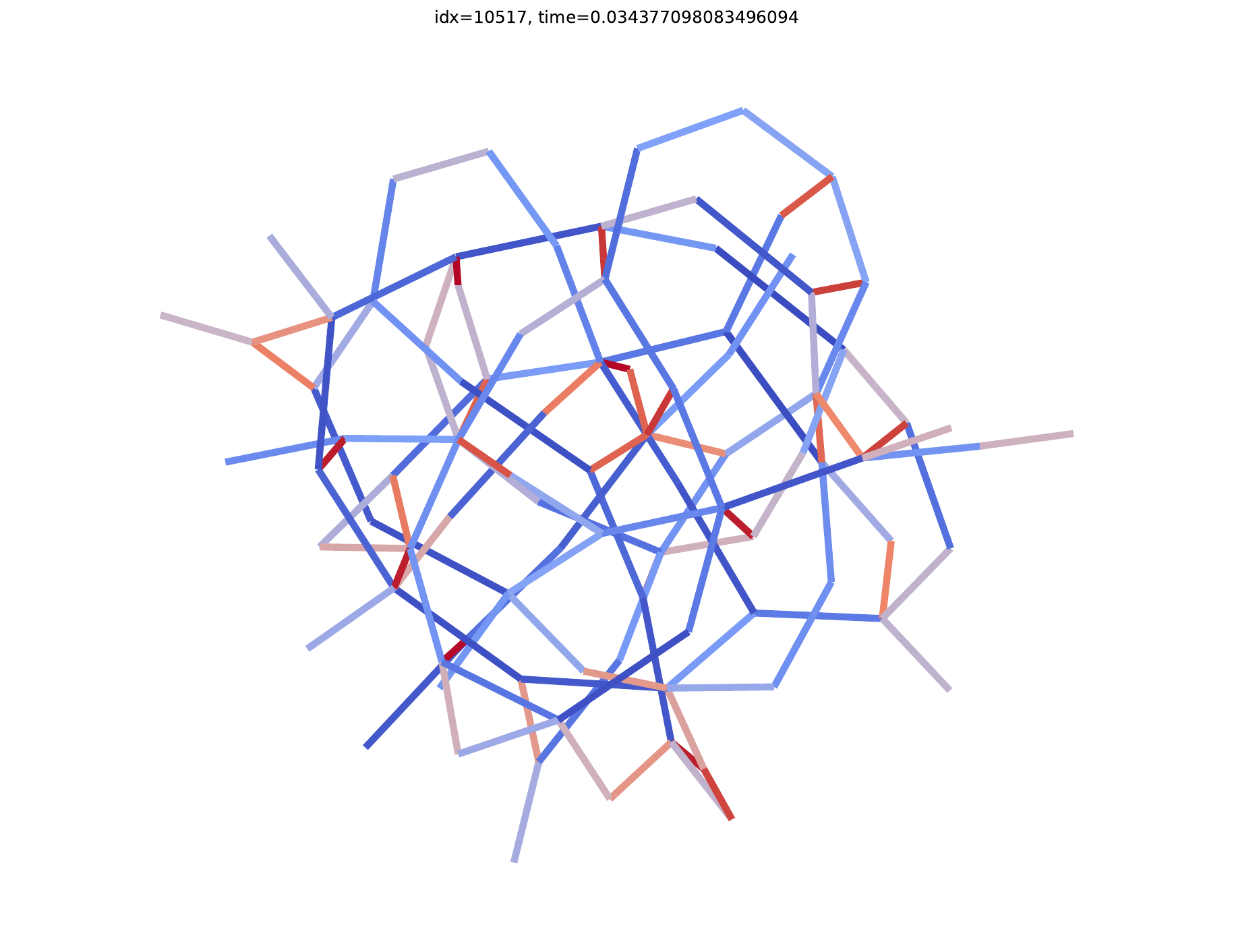} &
\imgcell{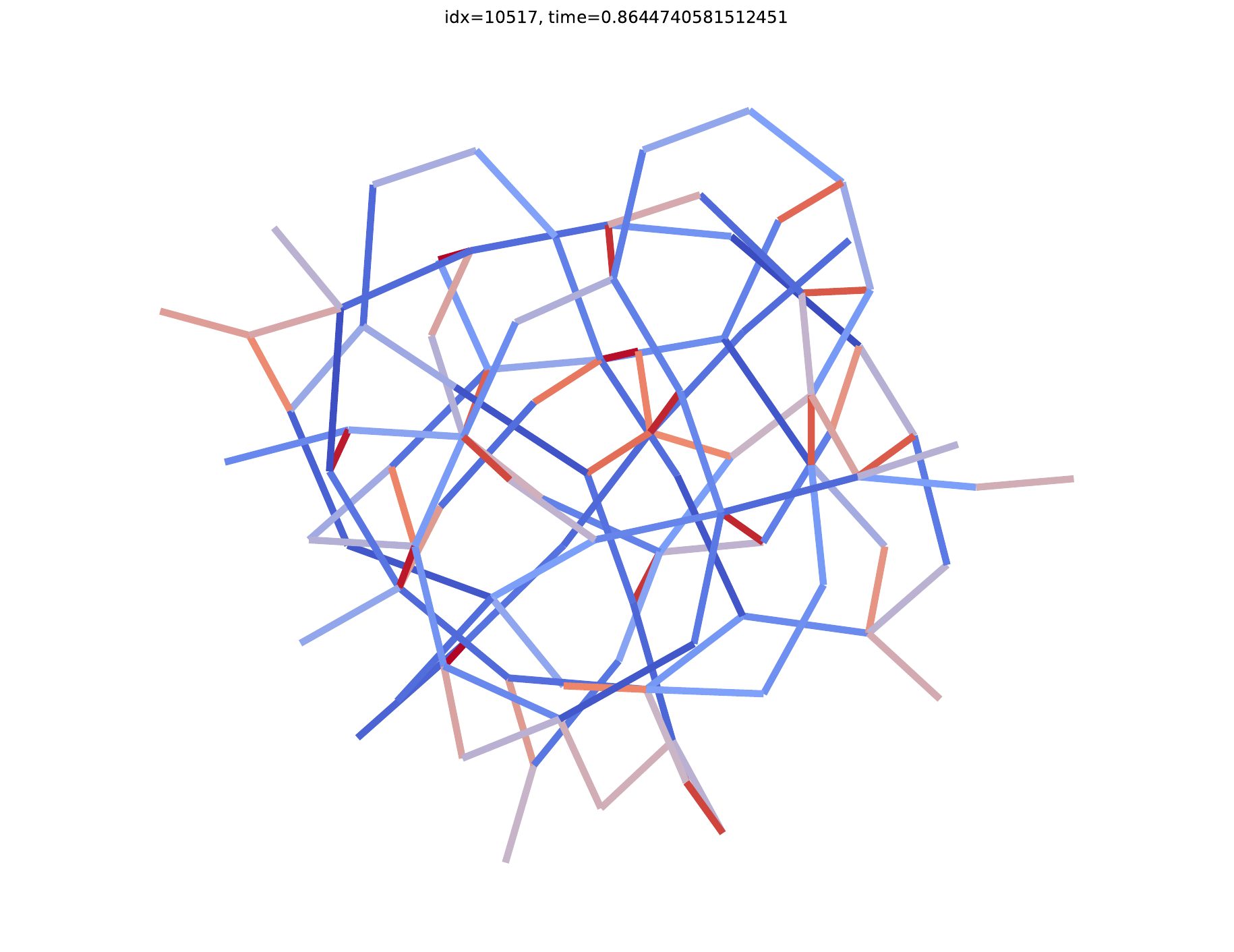} &
\imgcell{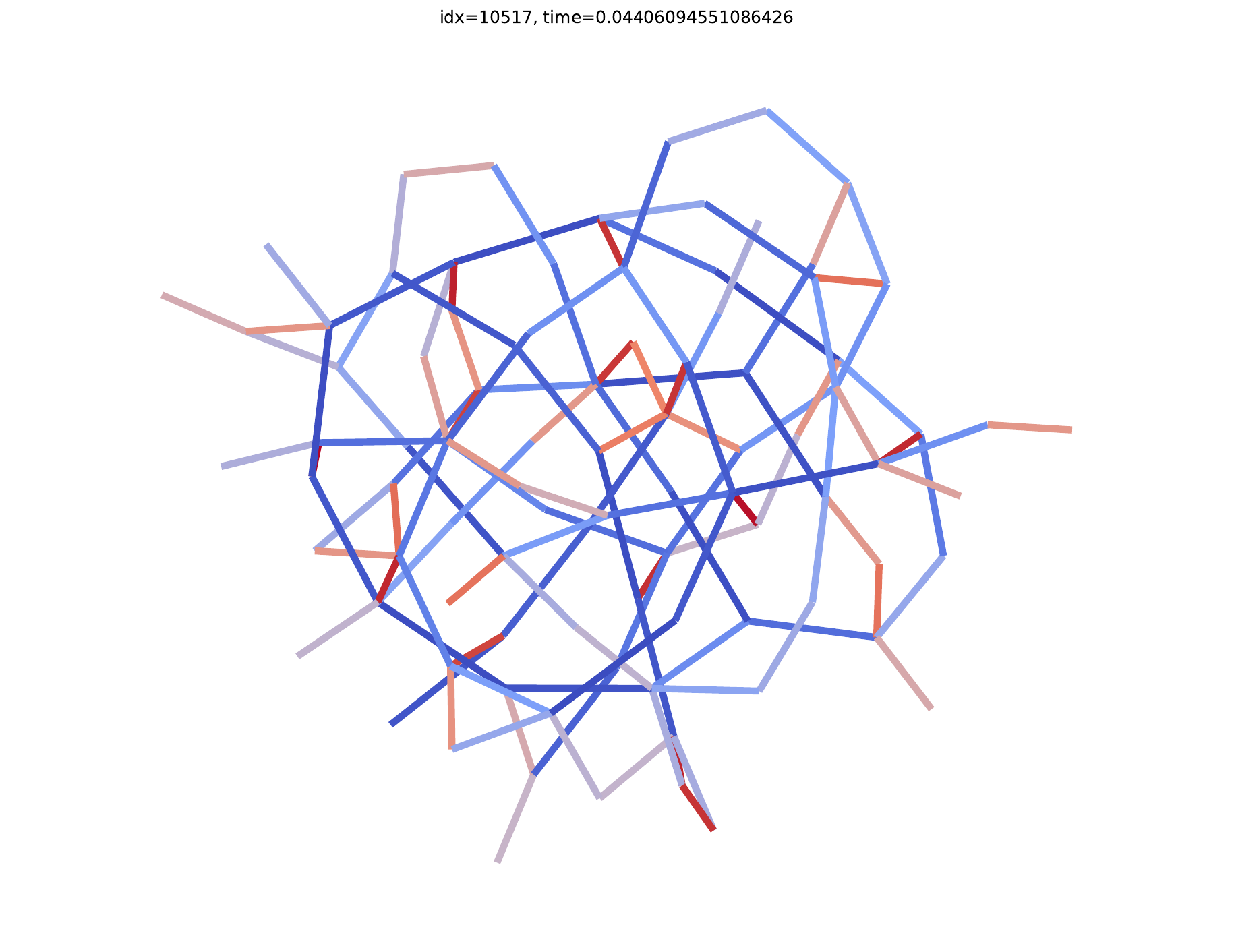} \\

&
t = 0.00s &
t = 1.70s &
t = 0.57s &
t = 0.05s &
t = 103.96s &
t = 0.05s &
t = 0.04s &
t = 0.04s &
t = 0.05s &
t = 0.03s &
t = 0.06s &
t = 0.04s \\

\makecell{\bfseries grafo10458.31\\N = 38\\M = 48} &
\imgcell{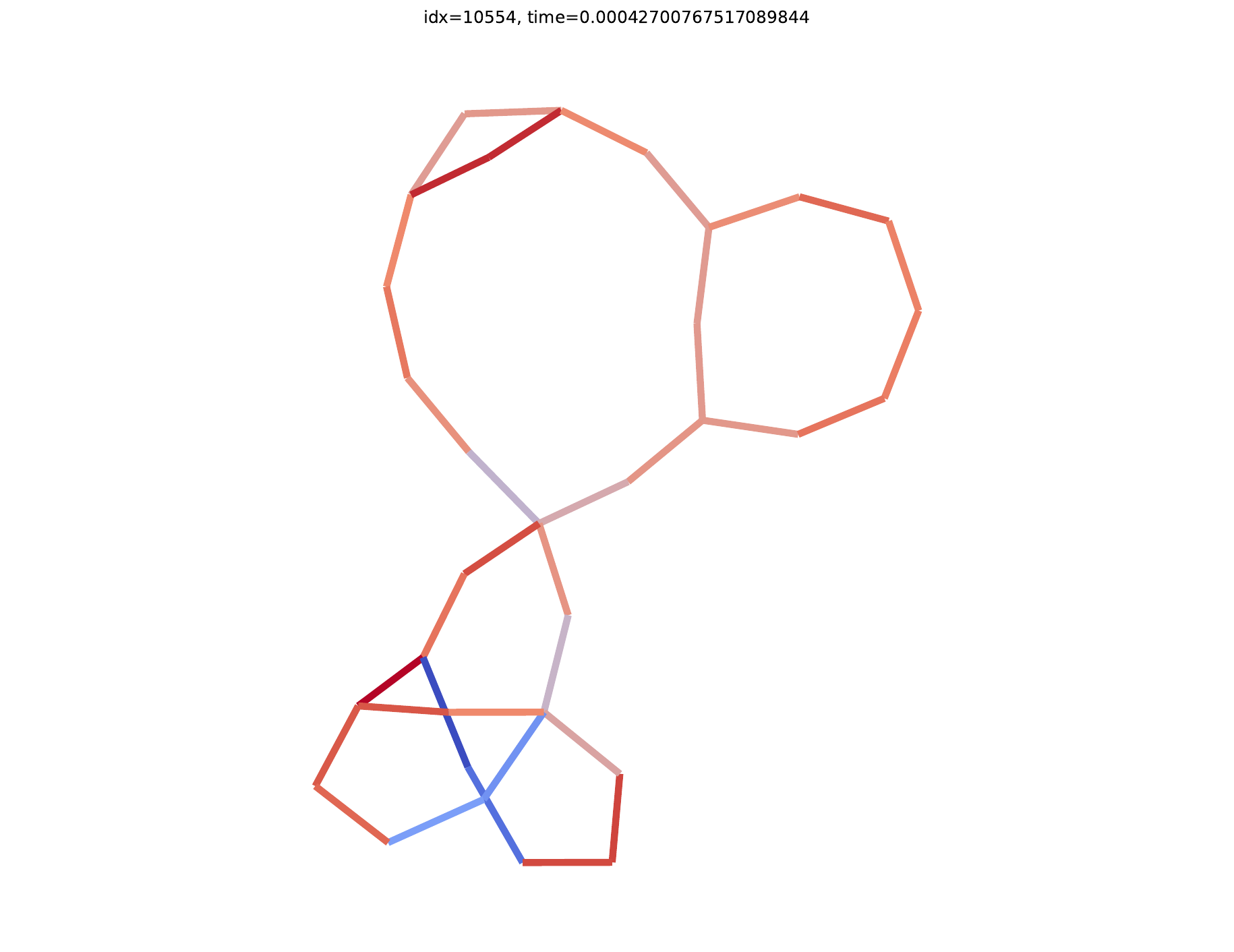} &
\imgcell{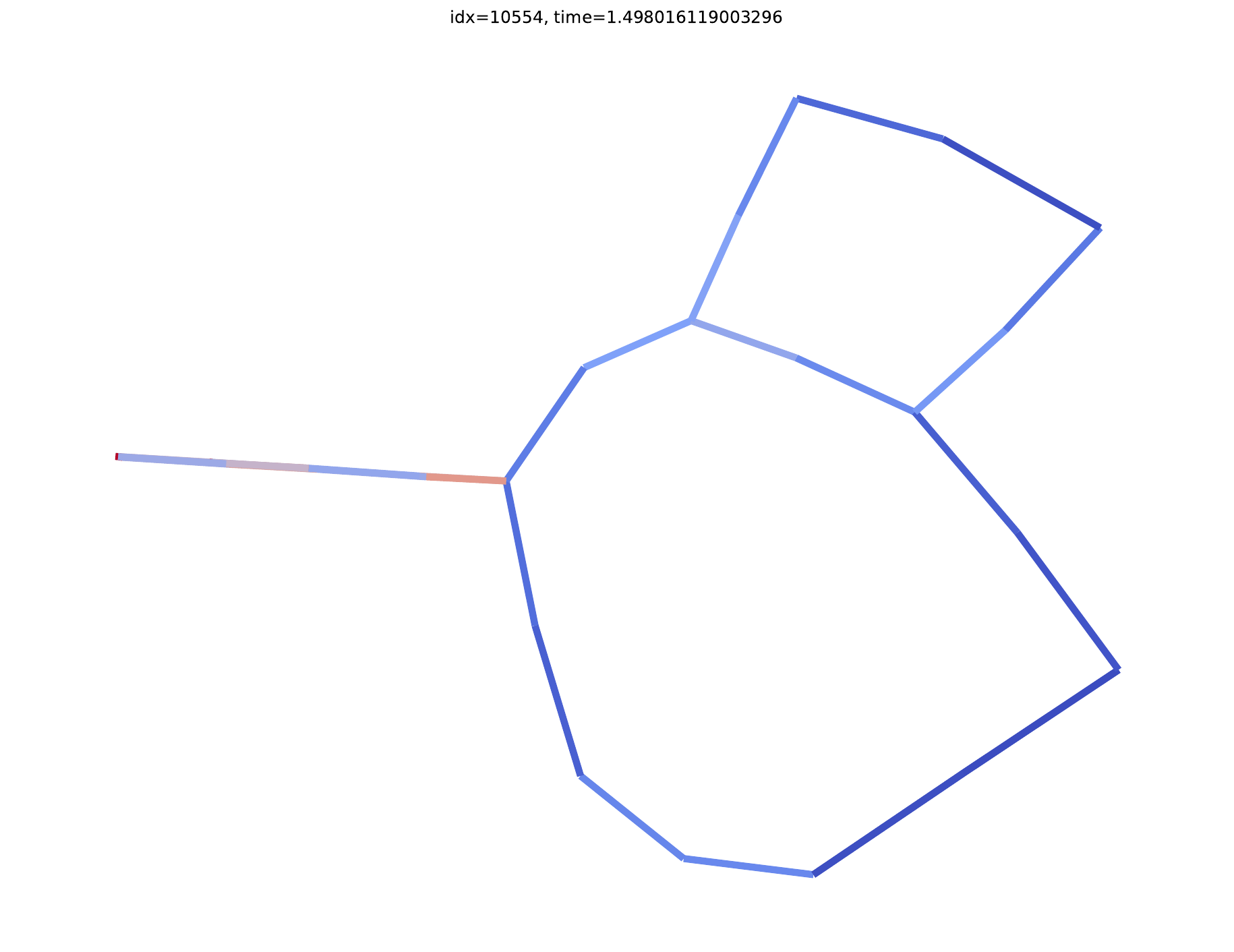} &
\imgcell{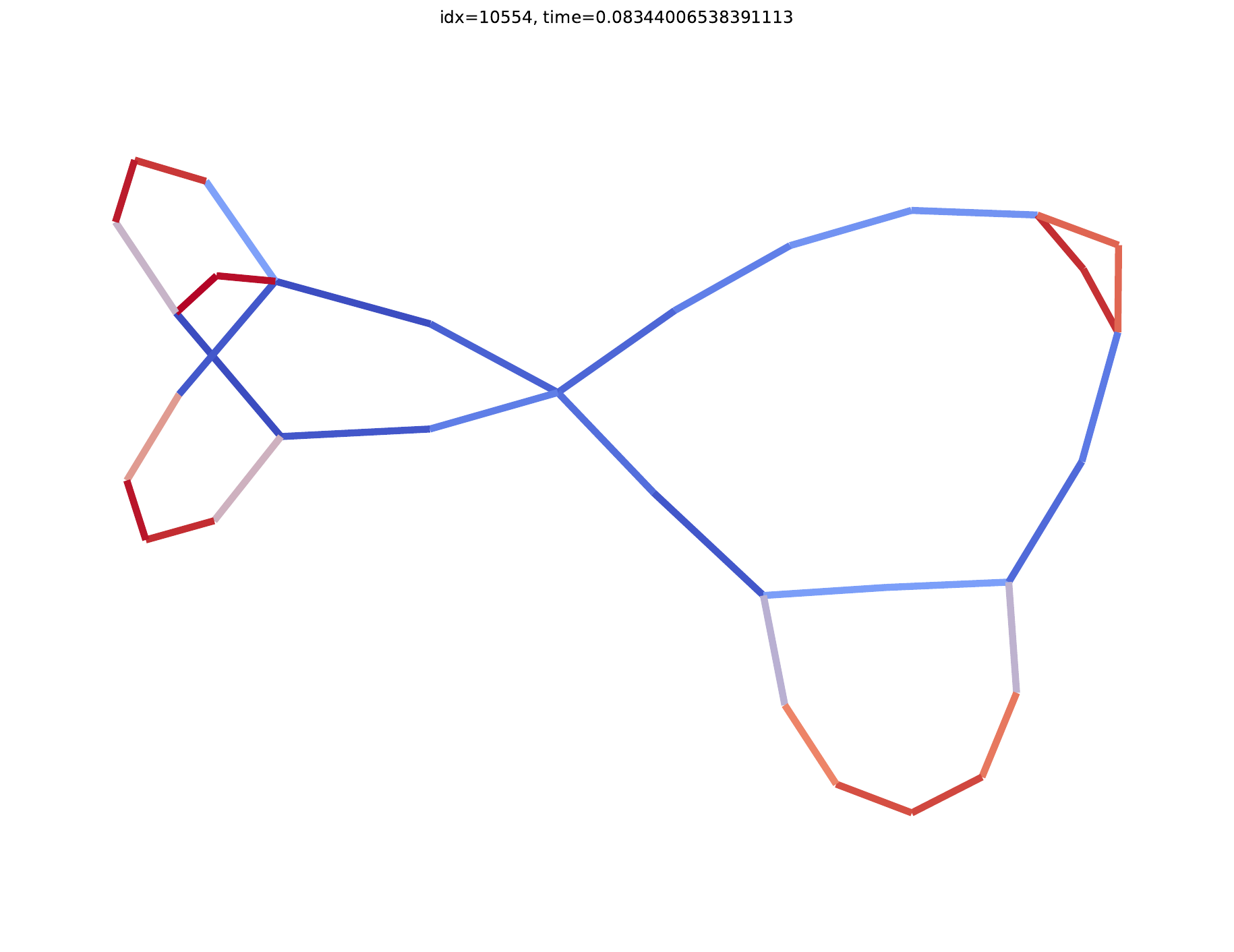} &
\imgcell{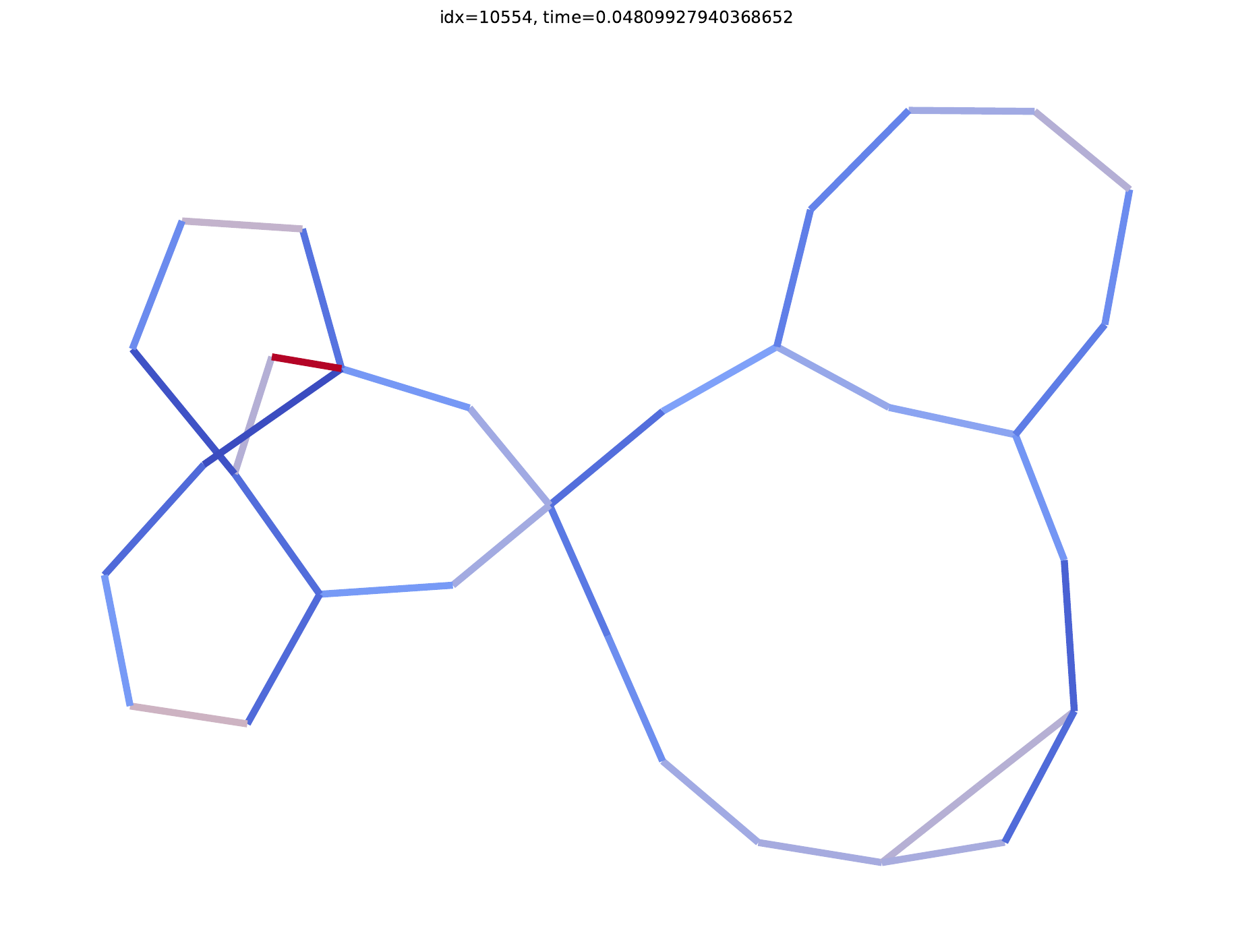} &
\imgcell{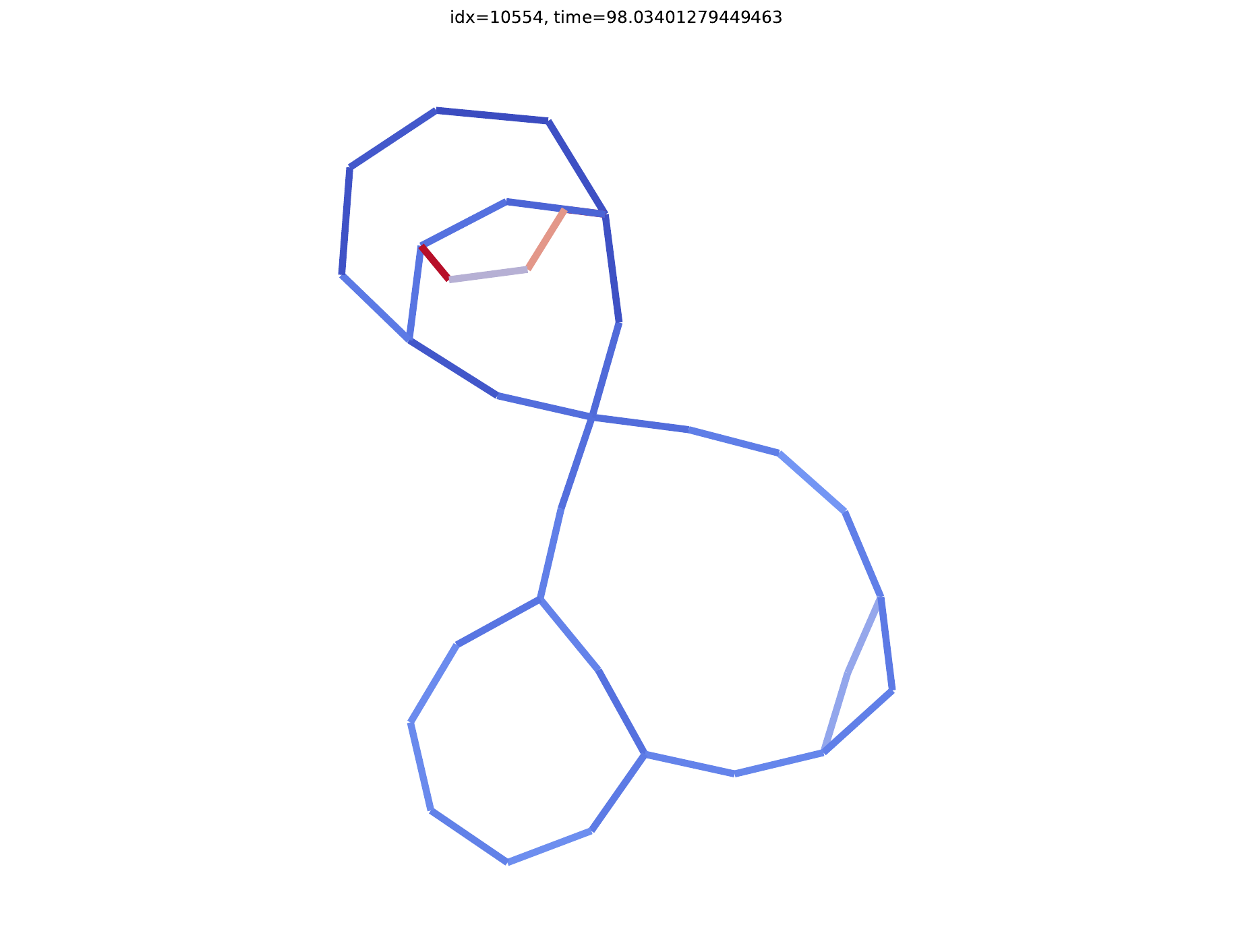} &
\imgcell{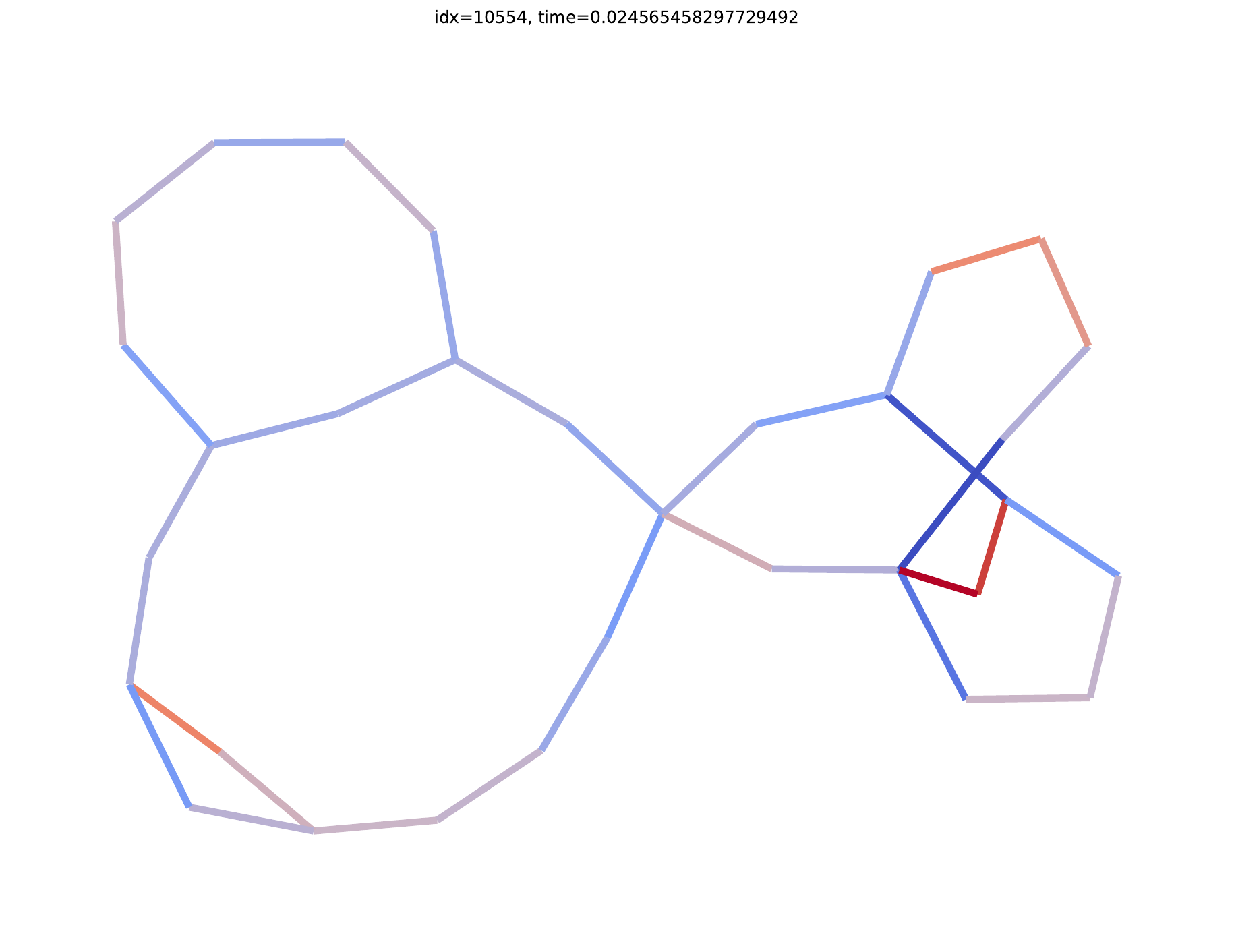} &
\imgcell{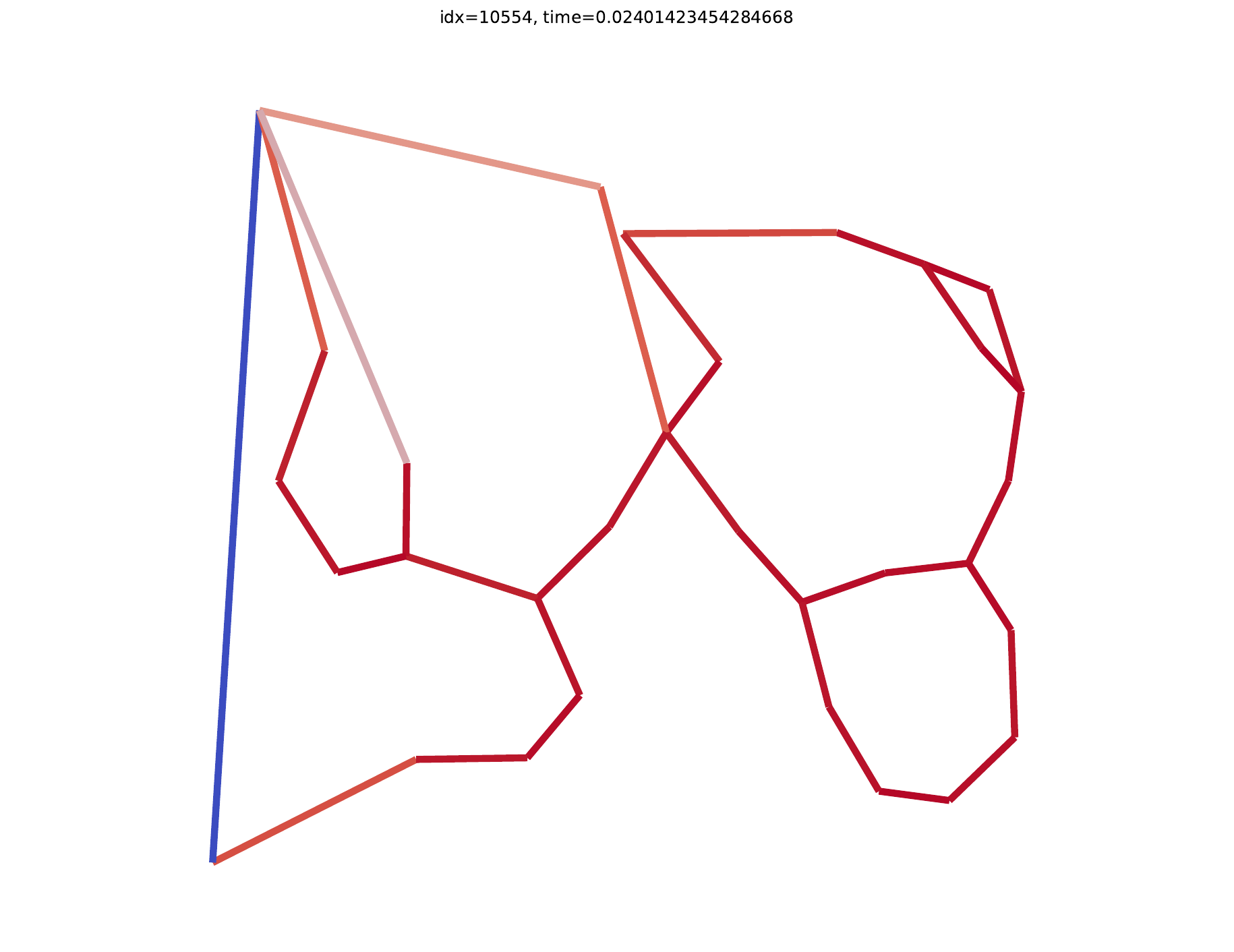} &
\imgcell{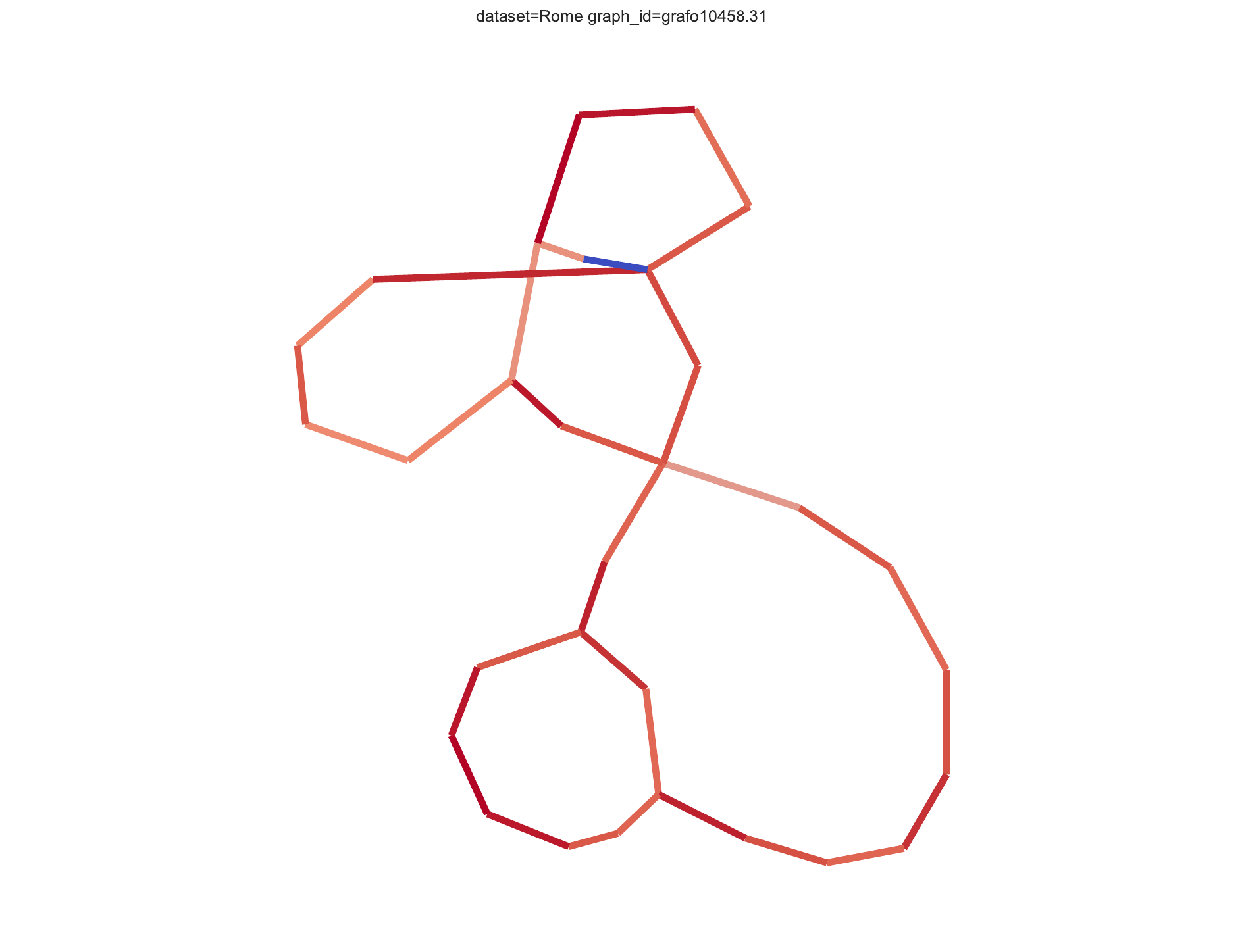} &
\imgcell{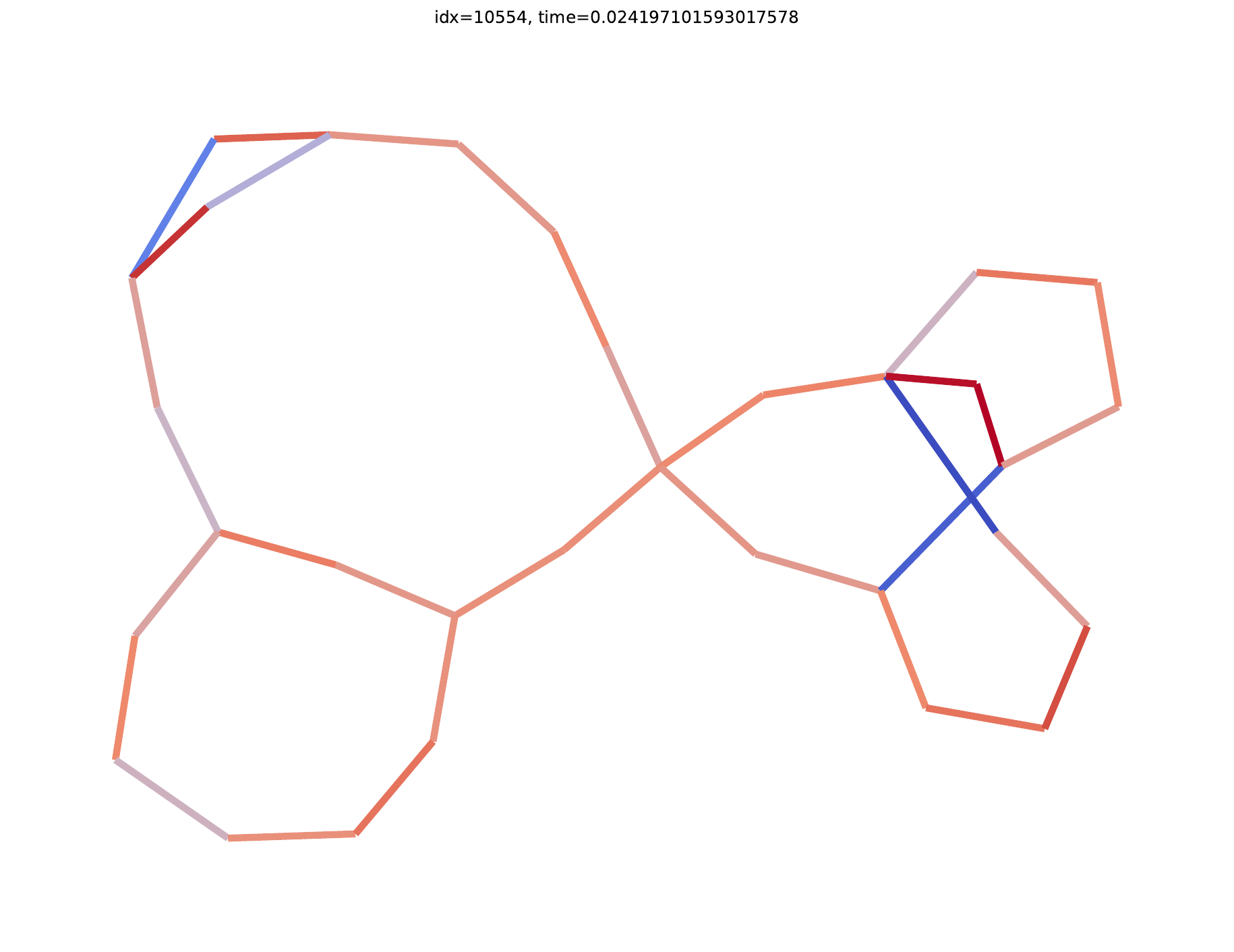} &
\imgcell{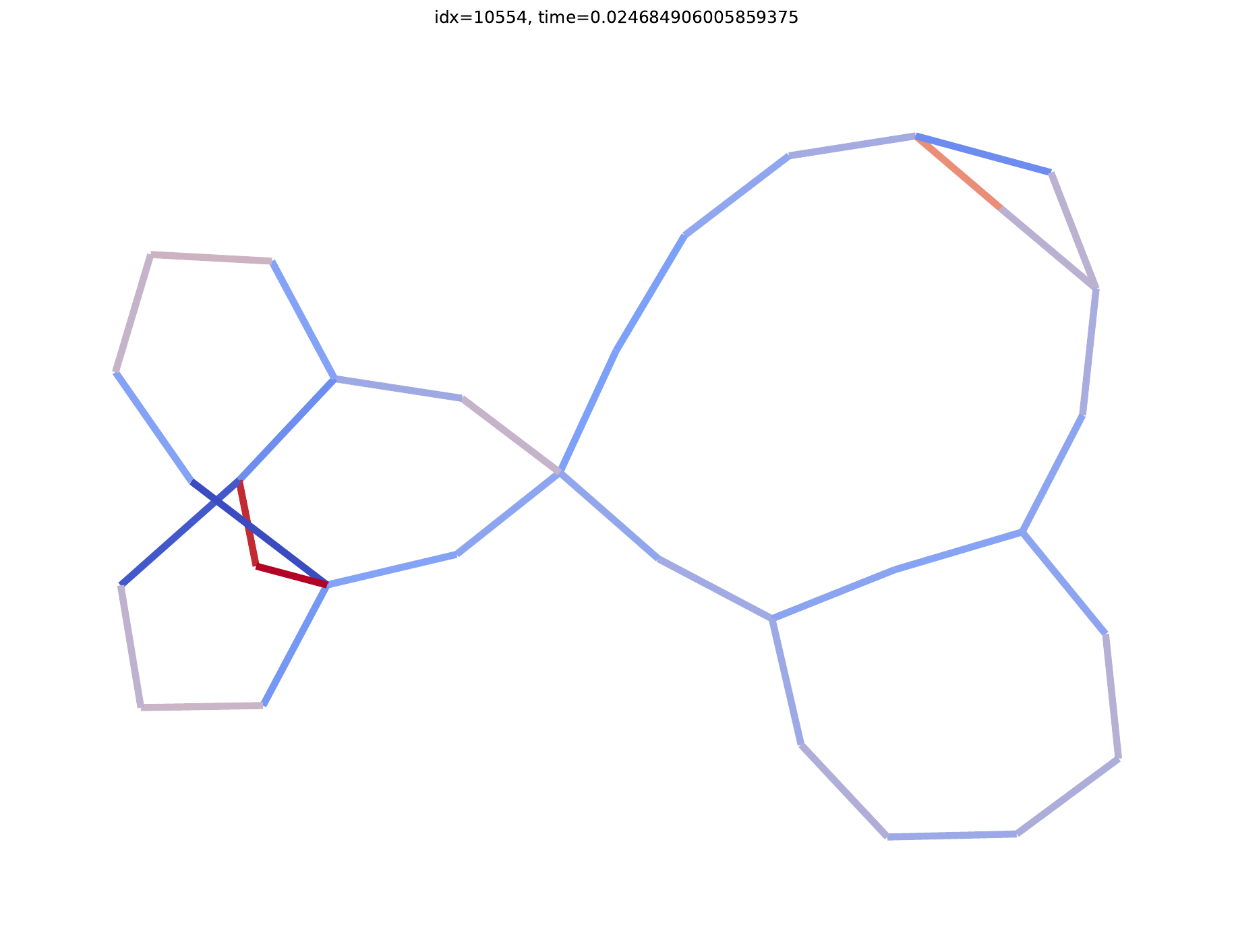} &
\imgcell{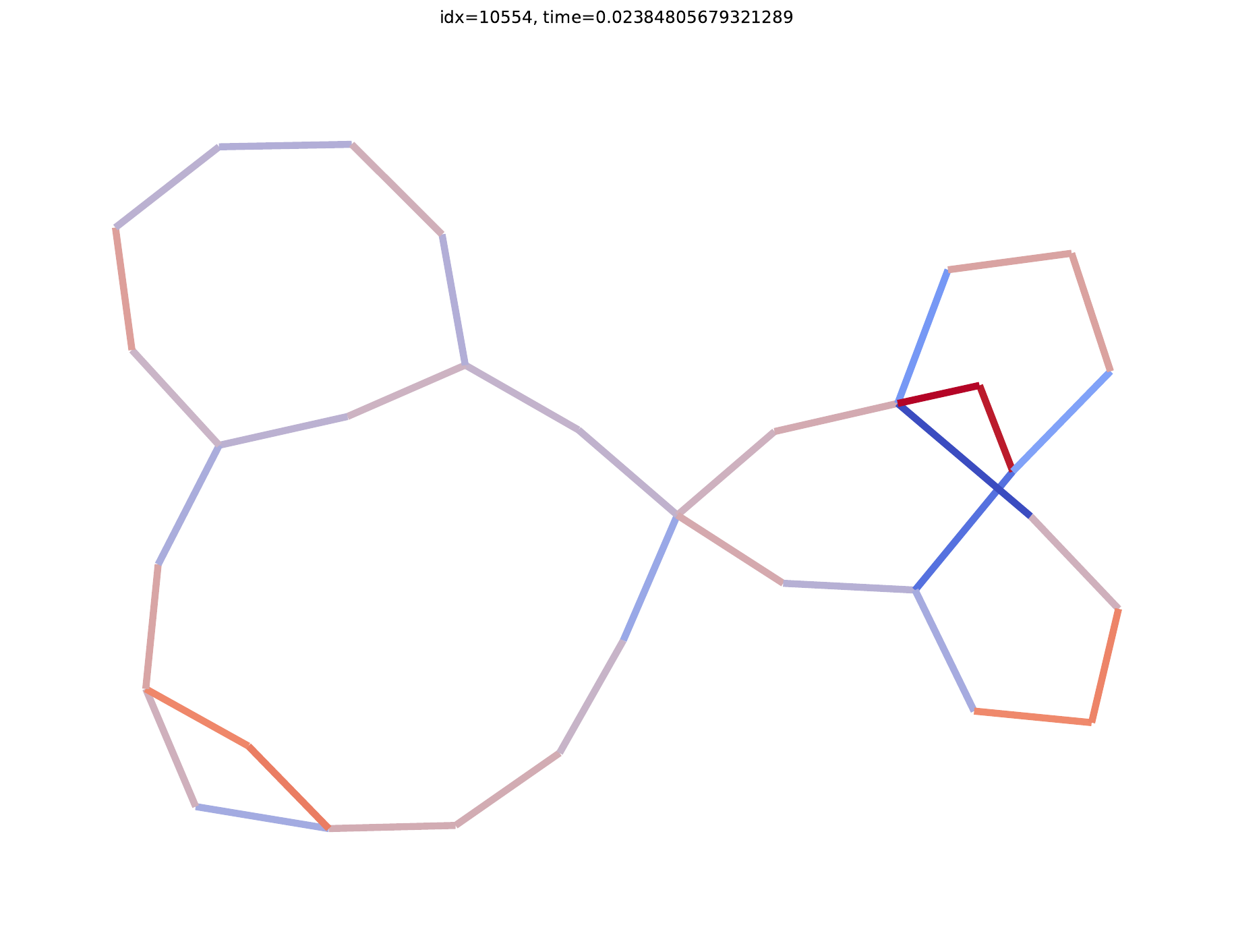} &
\imgcell{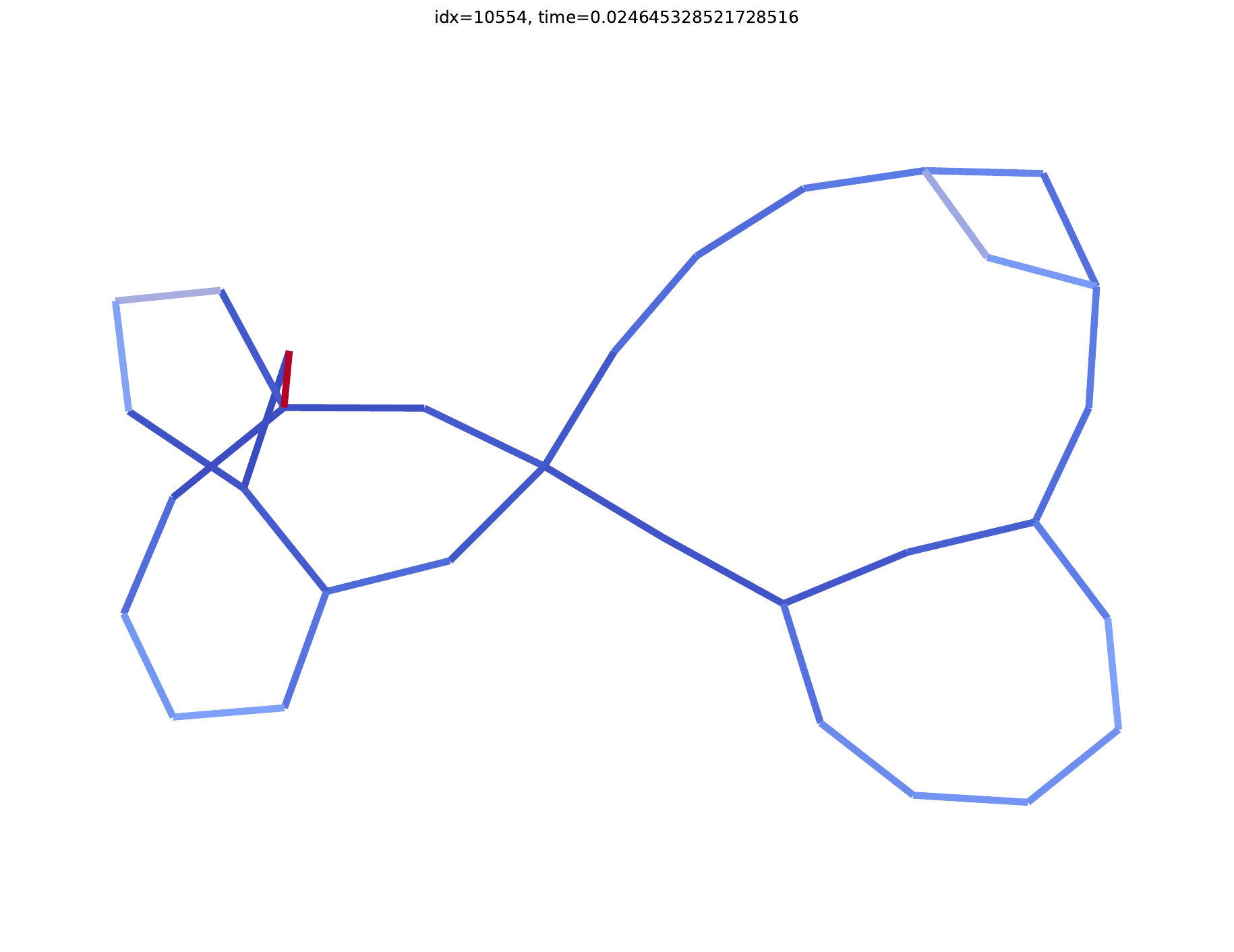} \\

&
t = 0.00s &
t = 1.50s &
t = 0.08s &
t = 0.05s &
t = 98.03s &
t = 0.02s &
t = 0.02s &
t = 0.02s &
t = 0.02s &
t = 0.02s &
t = 0.02s &
t = 0.02s \\

\end{tabular}
\captionof{figure}[]{The qualitative evaluation of 7 \modelName\ models by comparing with 5 competitive and representative benchmarks. All the graphs presented above are unseen during the training phase of \modelName. The name of the graphs with the number of nodes $N$ and the number of edges $M$ is presented in the row header. For each layout, the computation time $t$ (without including the pre-processing time) on the CPU is computed and reported in seconds. }
\label{fig:more-vis-result4}
\end{table*}

\begin{table*}[ht!]
\setlength{\tabcolsep}{0pt}
\renewcommand{\arraystretch}{0}
\fontsize{6}{6}\selectfont
\centering
\begin{tabular}{ c|ccccc|ccccccc }
    \bfseries{\thead{Graph}} & \multicolumn{5}{c|}{\thead{Benchmark Methods}} & \multicolumn{6}{c}{\thead{SmartGD}}\\
    & \bfseries{SGD2}
    & \bfseries{PMDS}
    & \bfseries{FA2}
    & \bfseries{DeepGD}
    & \makecell{\bfseries GD2\\\relax[Stress+Xing]}
    & \makecell{\bfseries SmartGD\\\relax[Stress]}
    & \makecell{\bfseries SmartGD\\\relax[Xing]}
    & \makecell{\bfseries SmartGD\\\relax[Shape]}
    & \makecell{\bfseries SmartGD\\\relax[XAngle]}
    & \makecell{\bfseries SmartGD\\\relax[Stress+Xing]}
    & \makecell{\bfseries SmartGD\\\relax[Stress+XAngle]}
    & \makecell{\bfseries SmartGD\\\relax[7-Aesthetics]}
    \rule[-1ex]{0pt}{0ex} \\ \hline

\makecell{\bfseries grafo2401.17\\N = 97\\M = 136} &
\imgcell{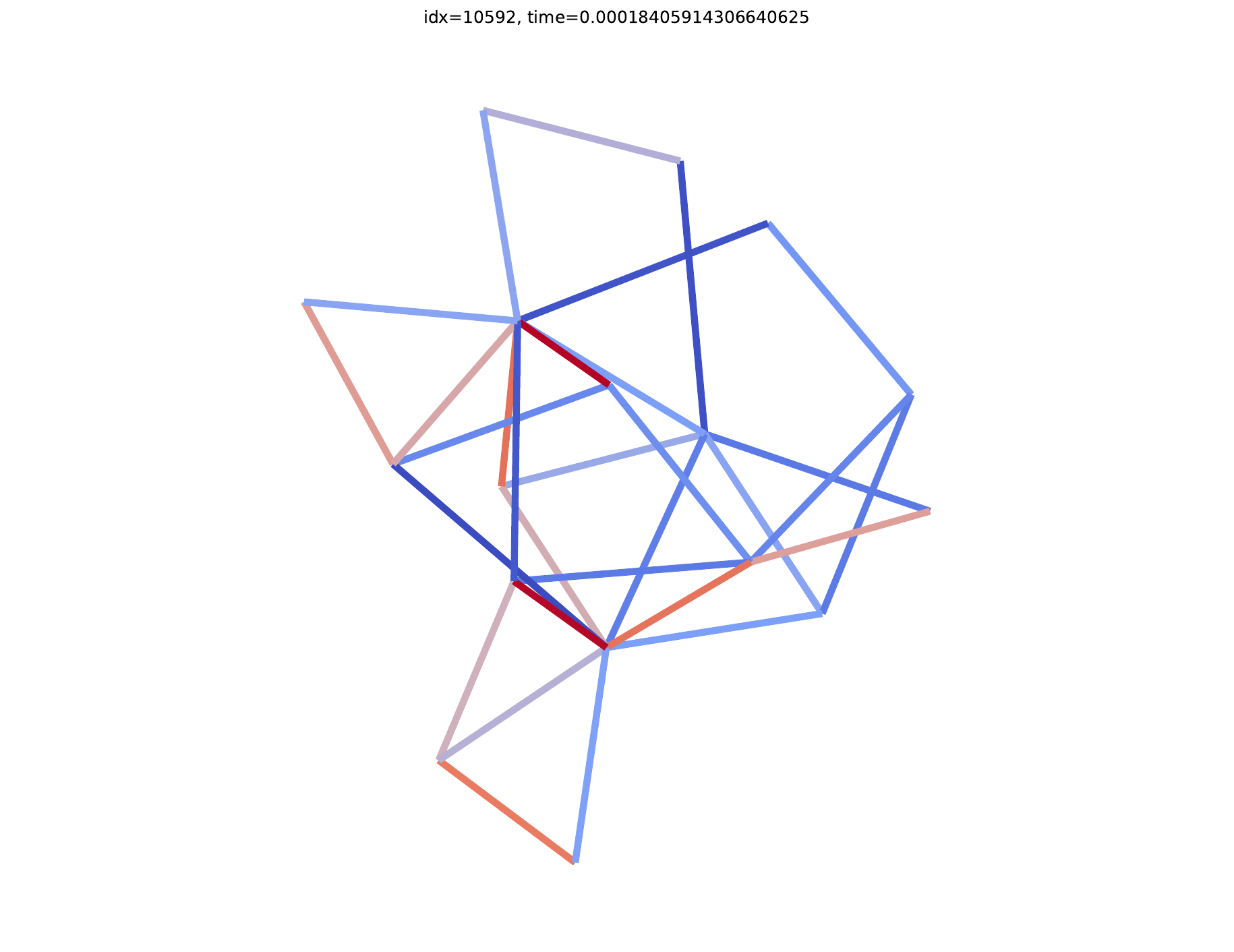} &
\imgcell{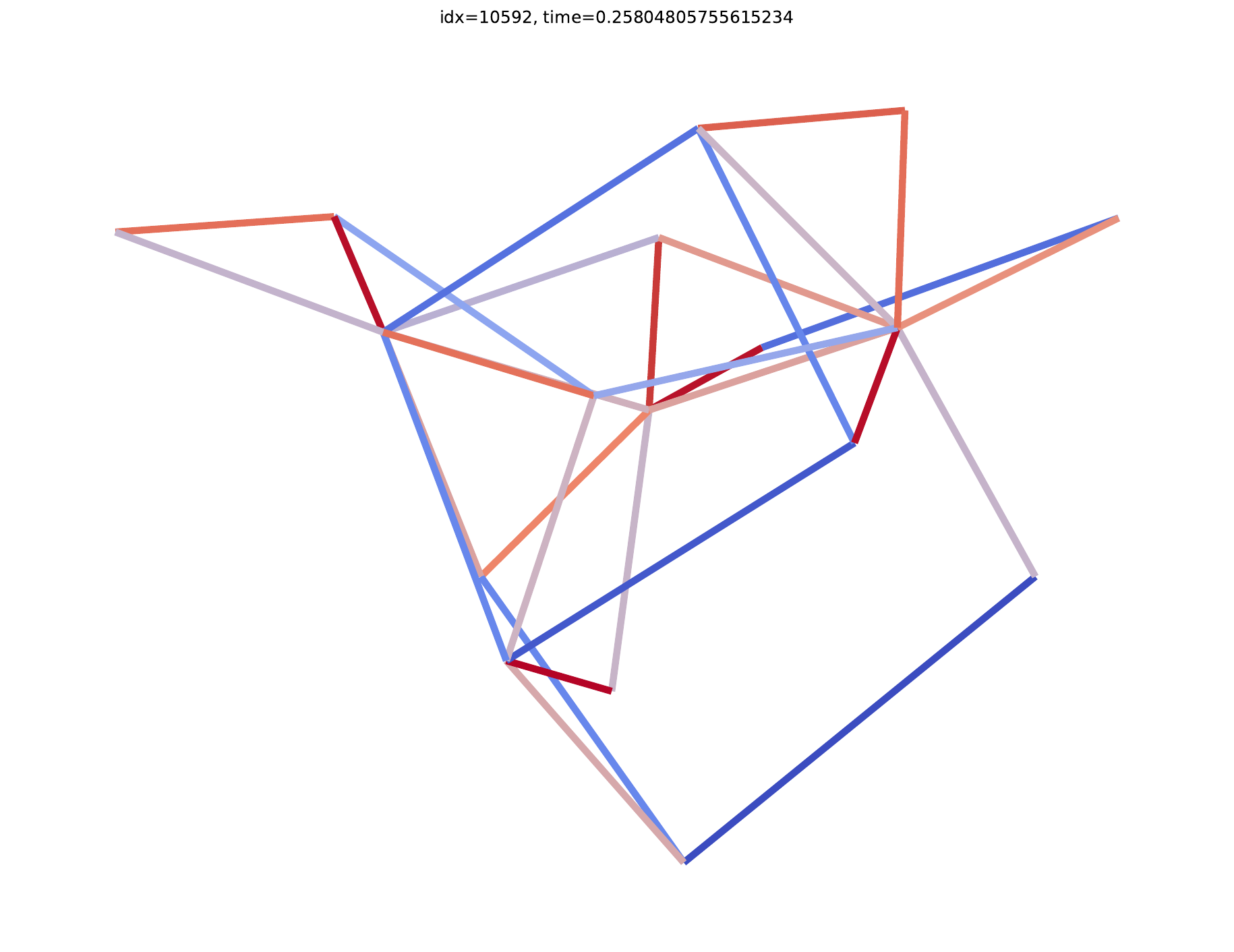} &
\imgcell{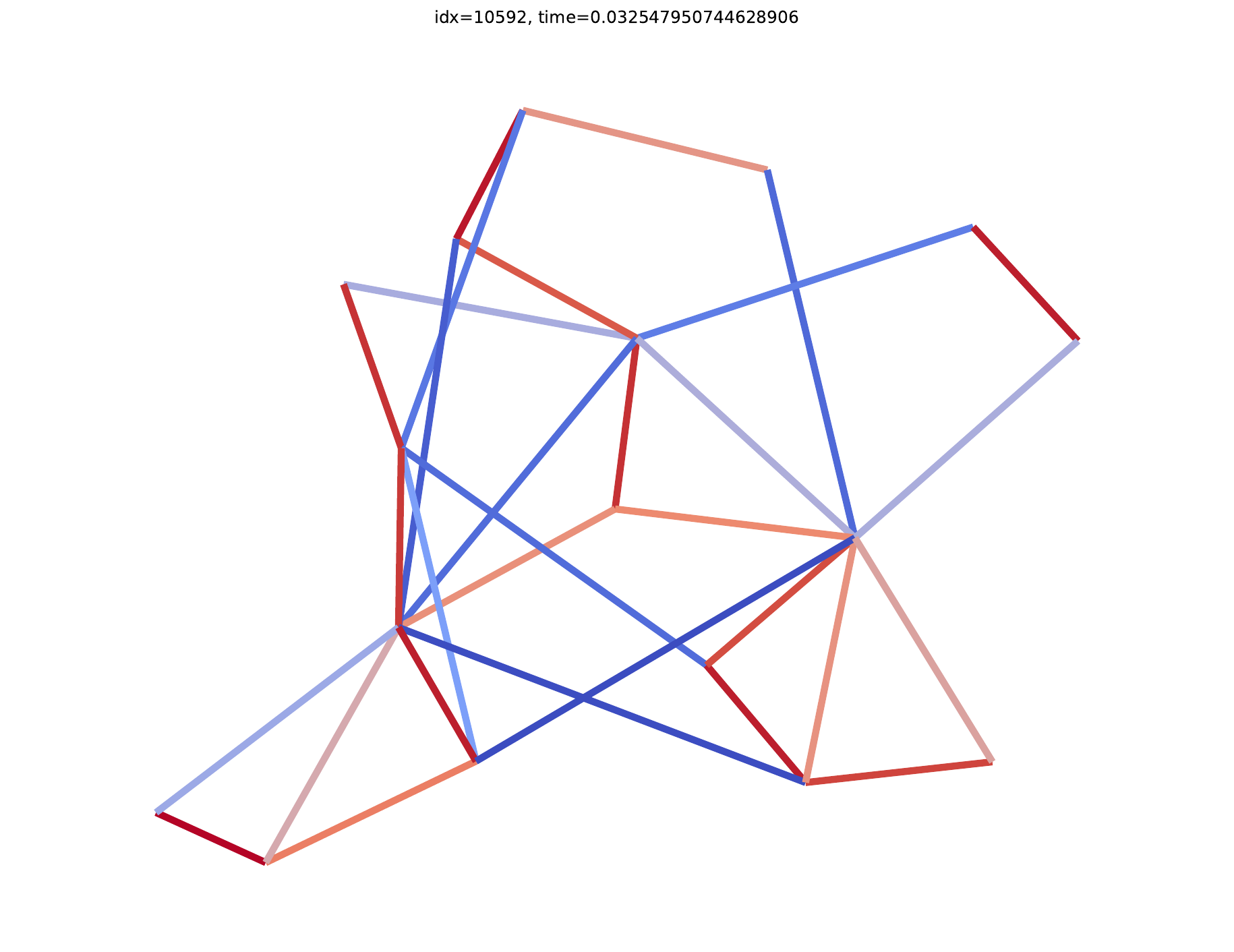} &
\imgcell{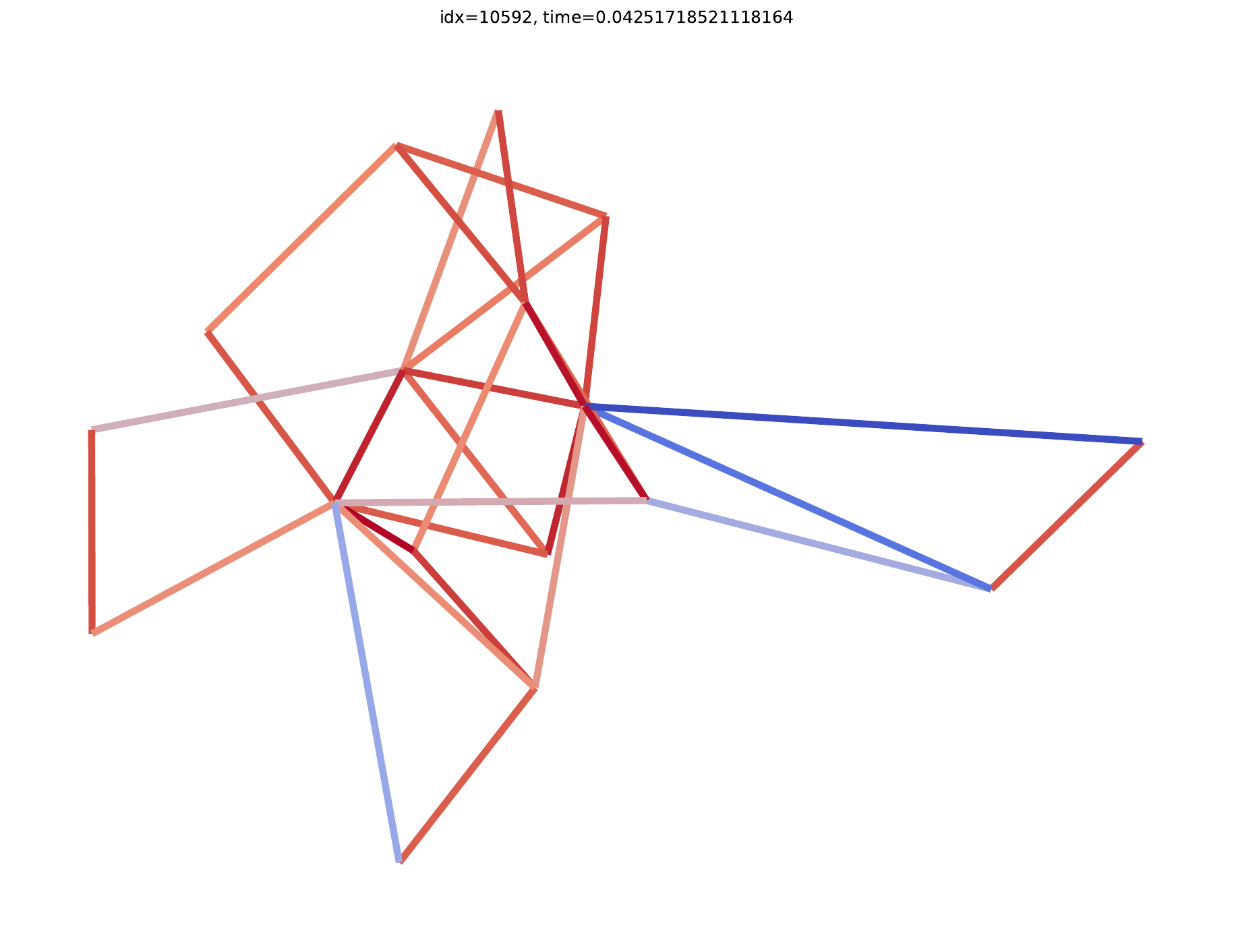} &
\imgcell{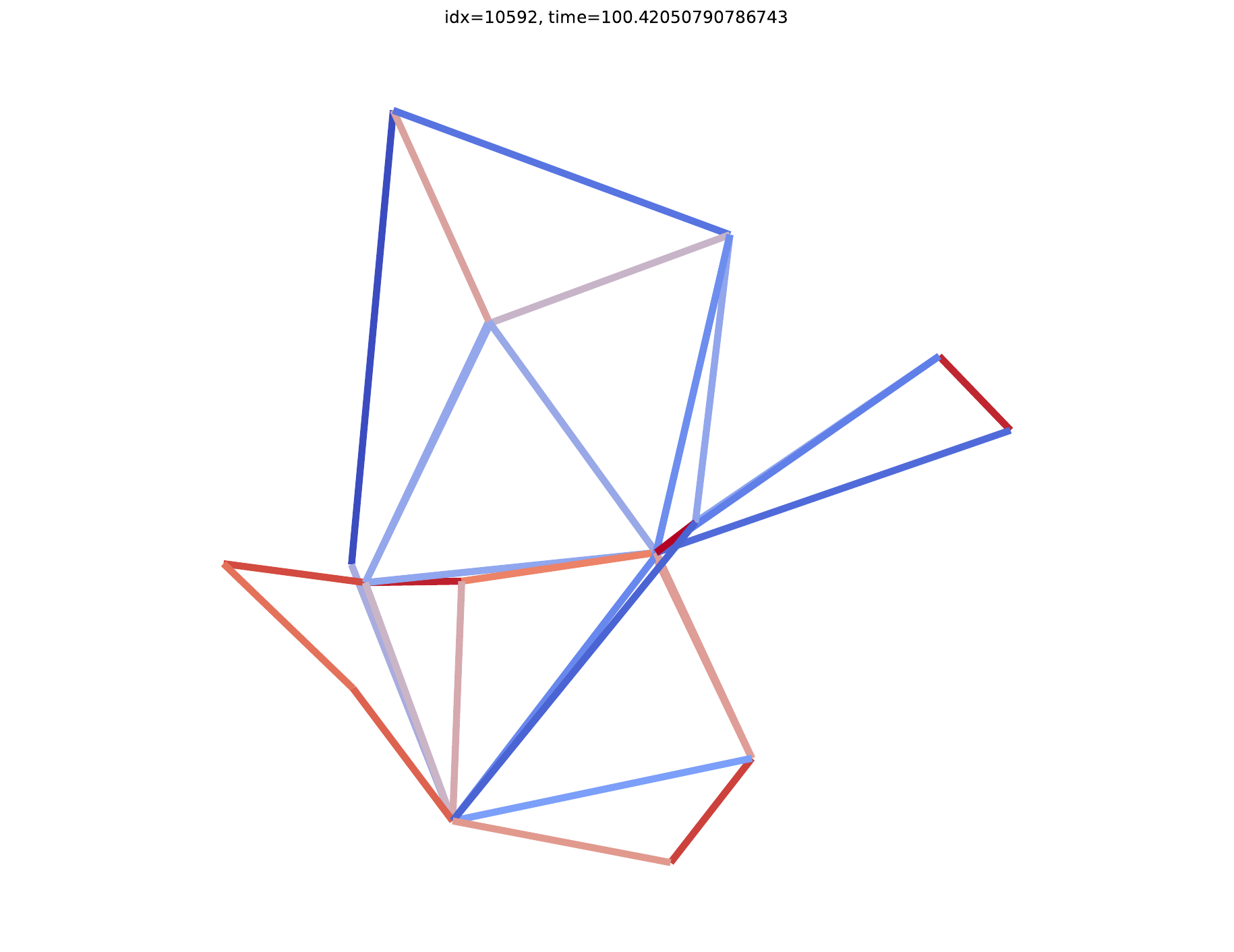} &
\imgcell{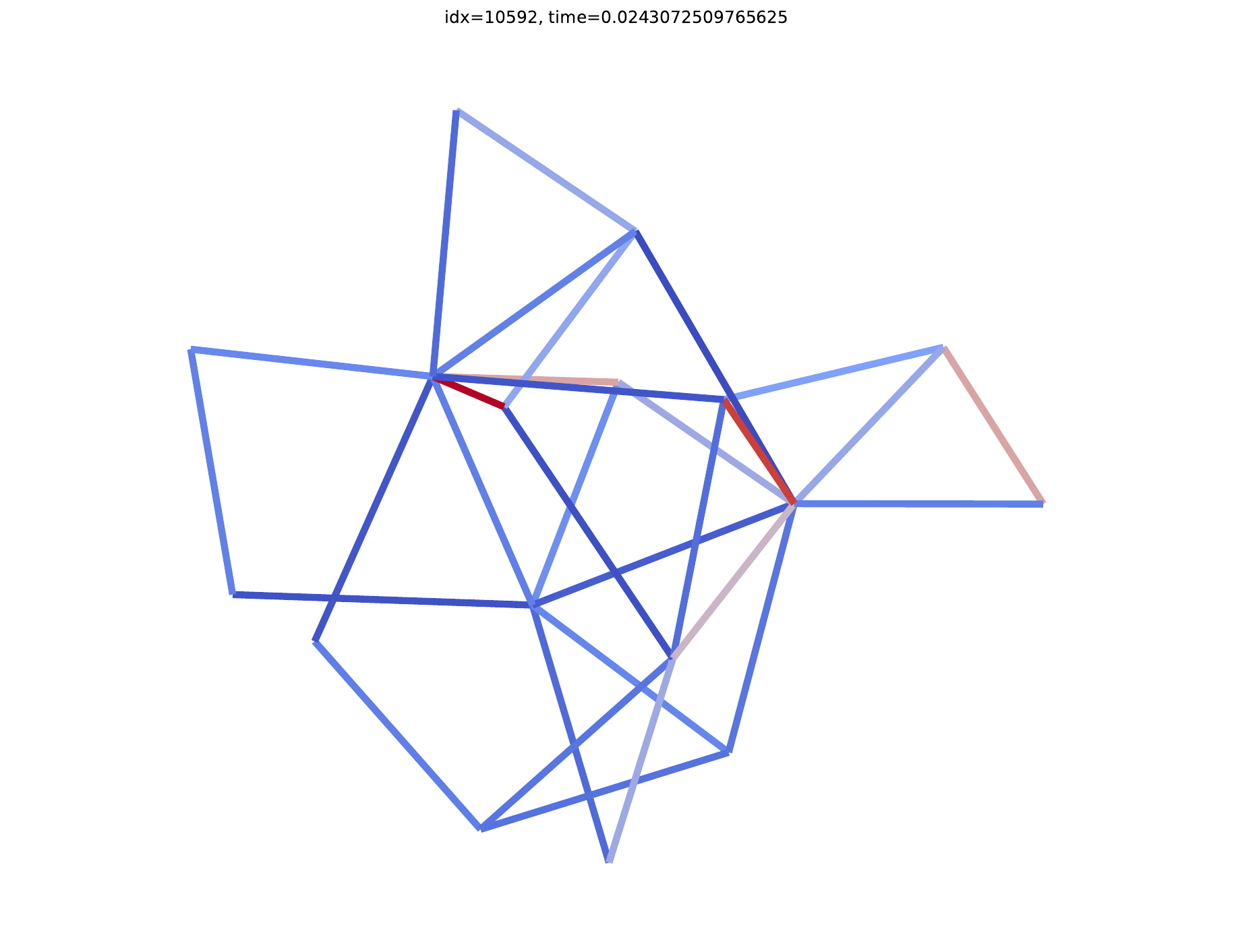} &
\imgcell{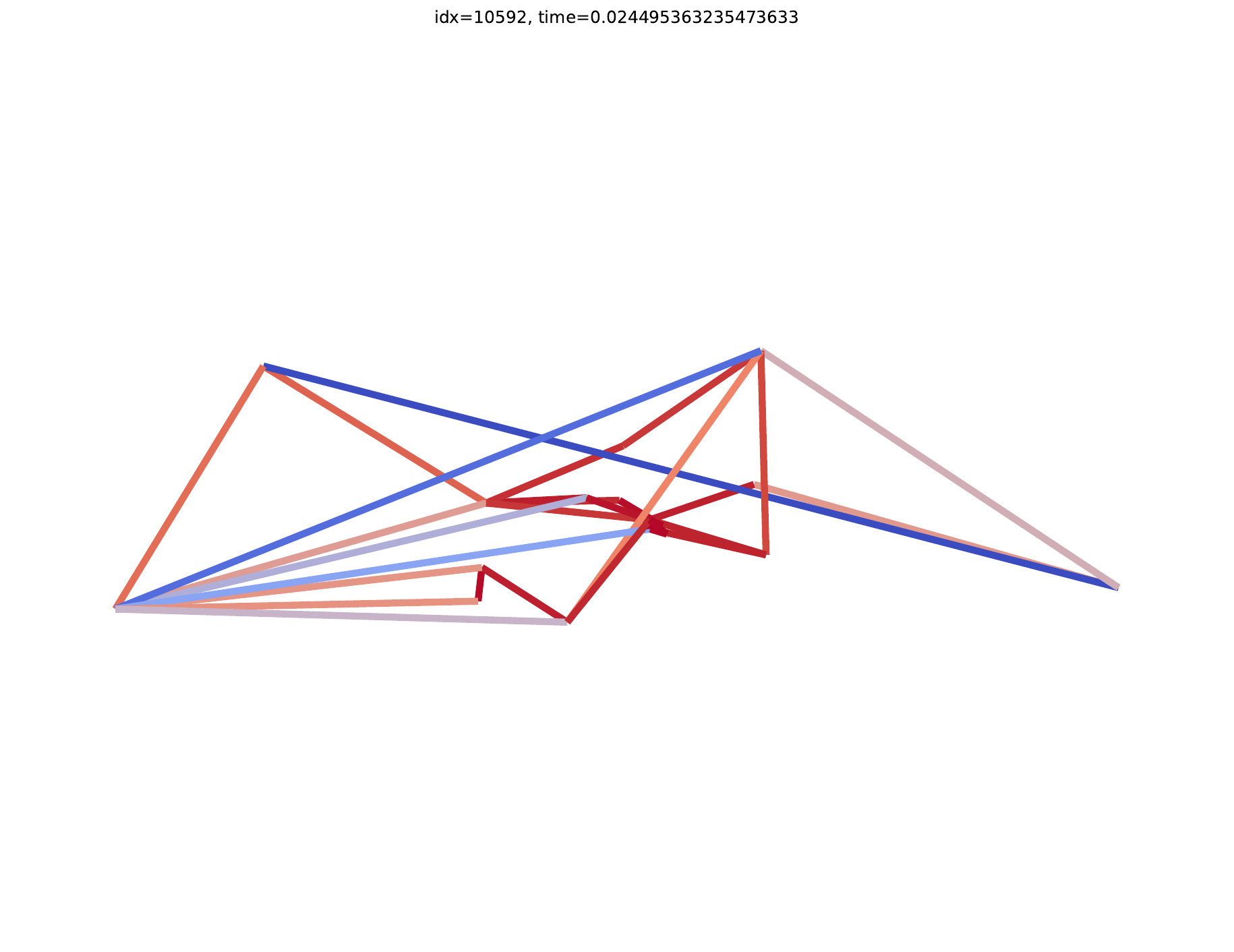} &
\imgcell{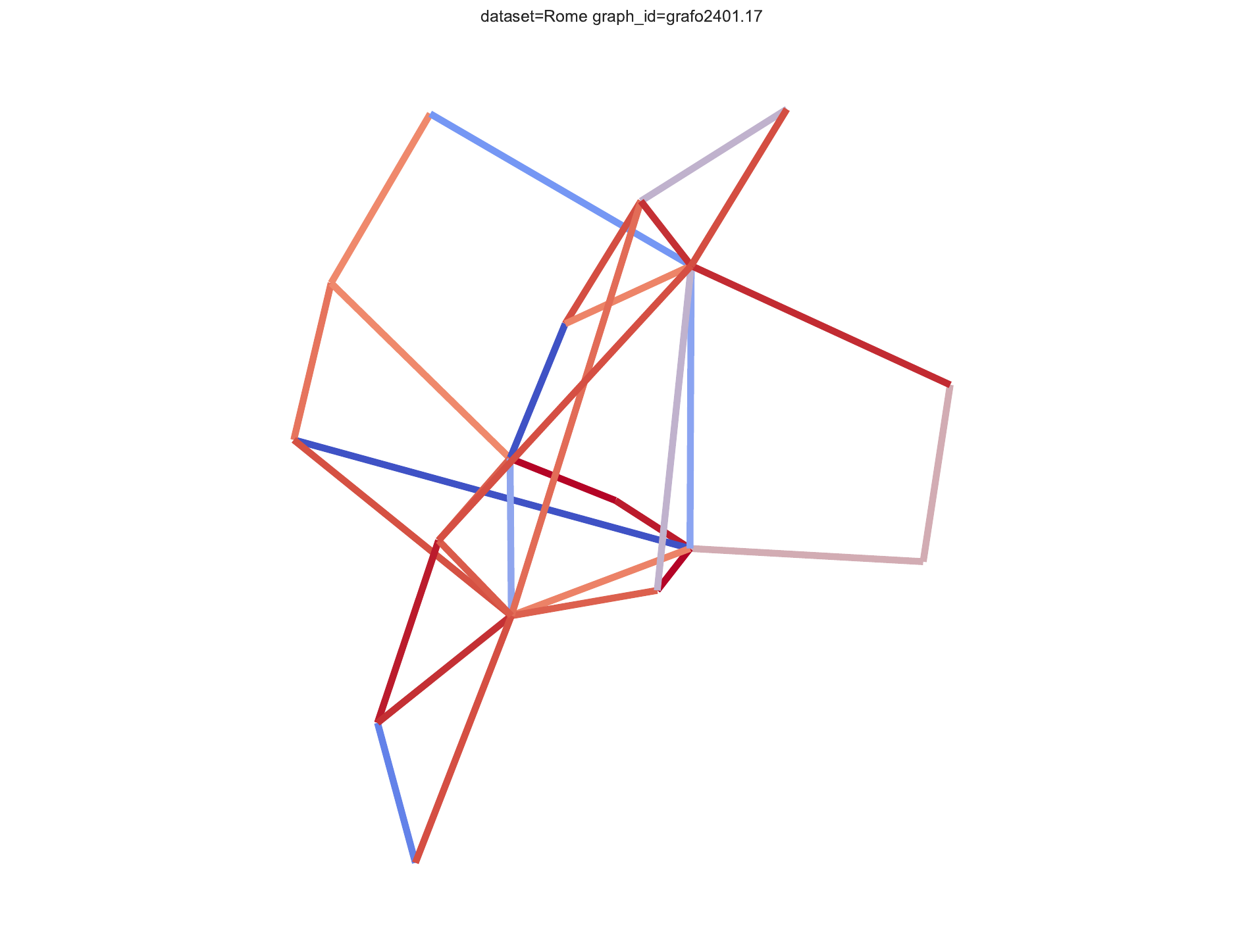} &
\imgcell{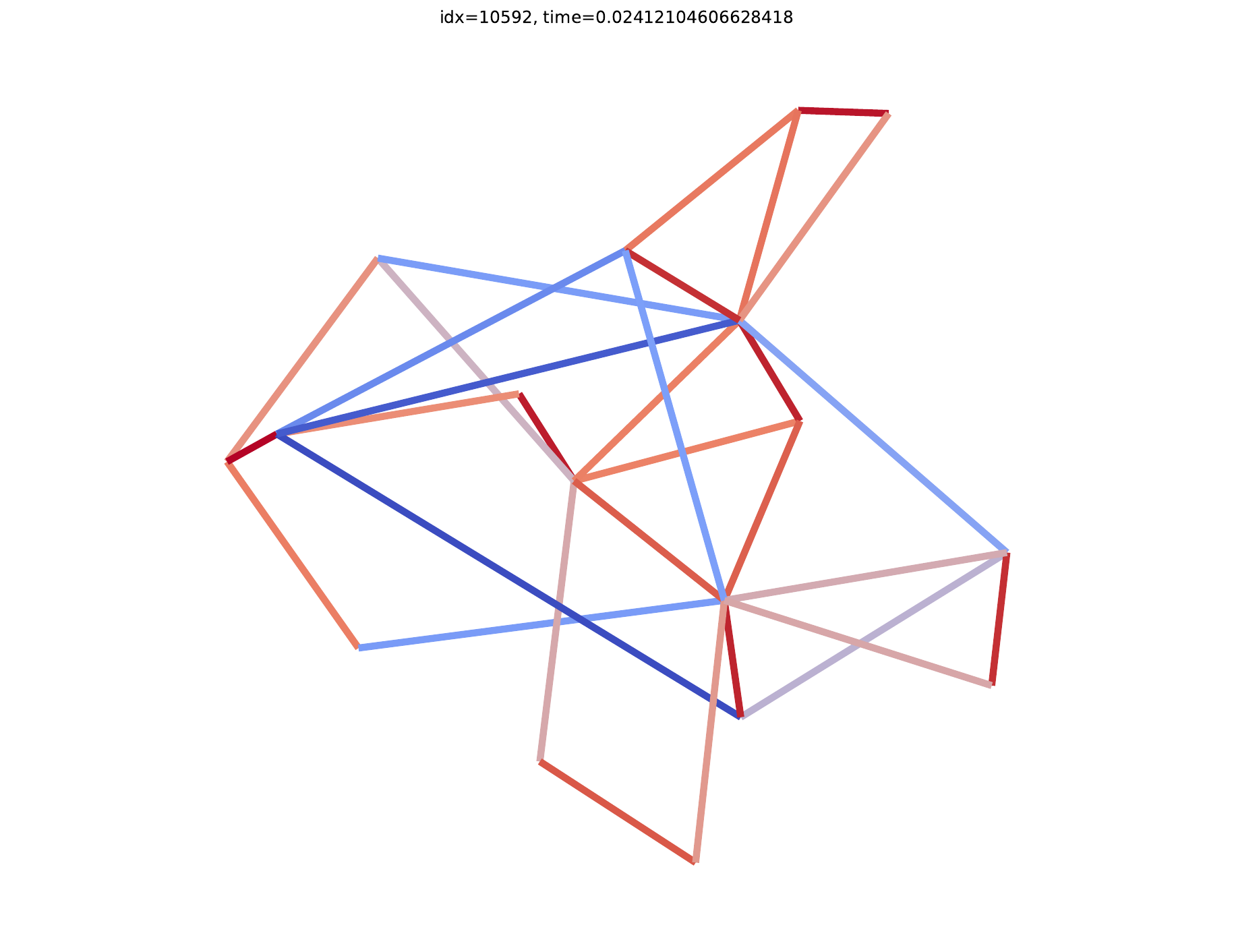} &
\imgcell{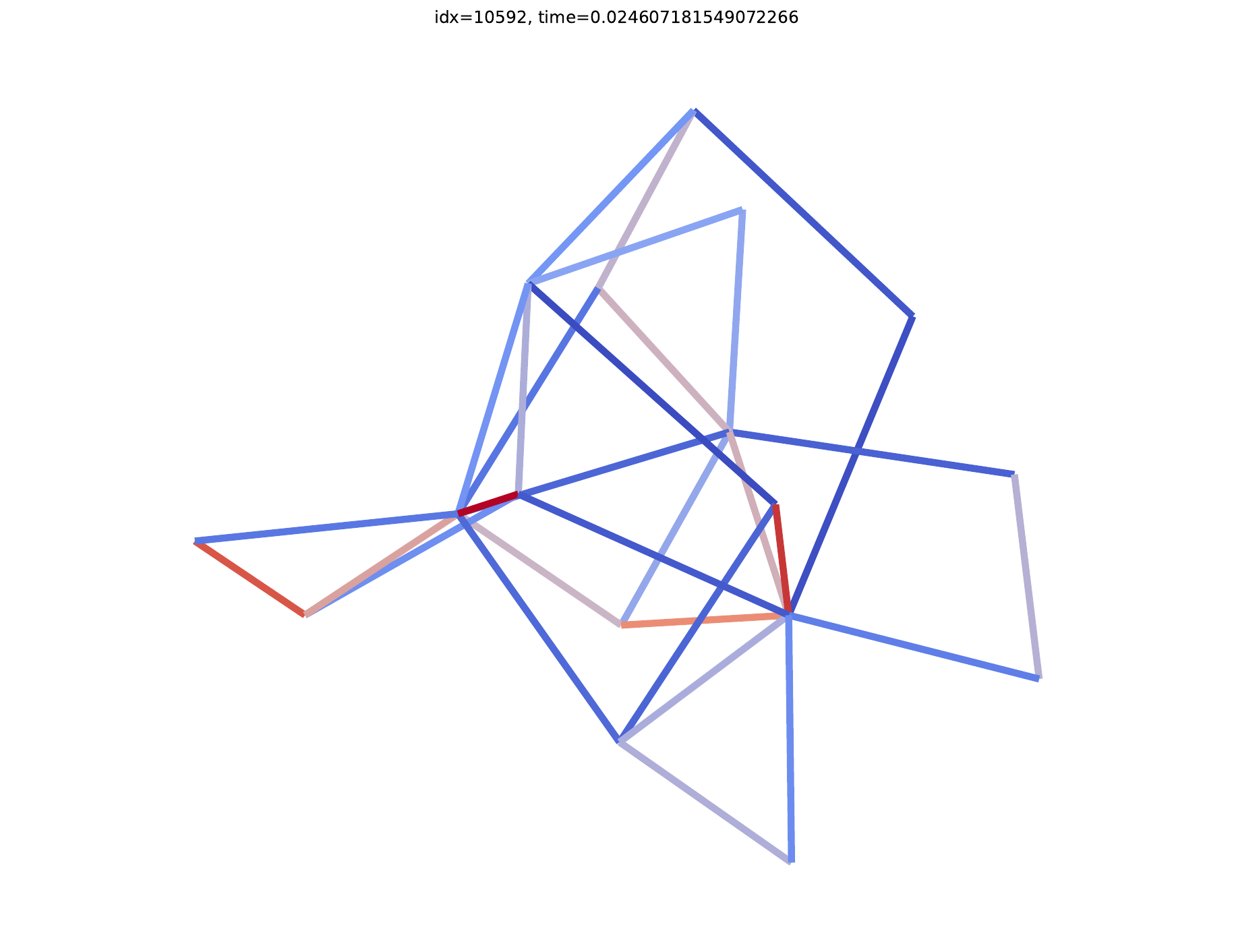} &
\imgcell{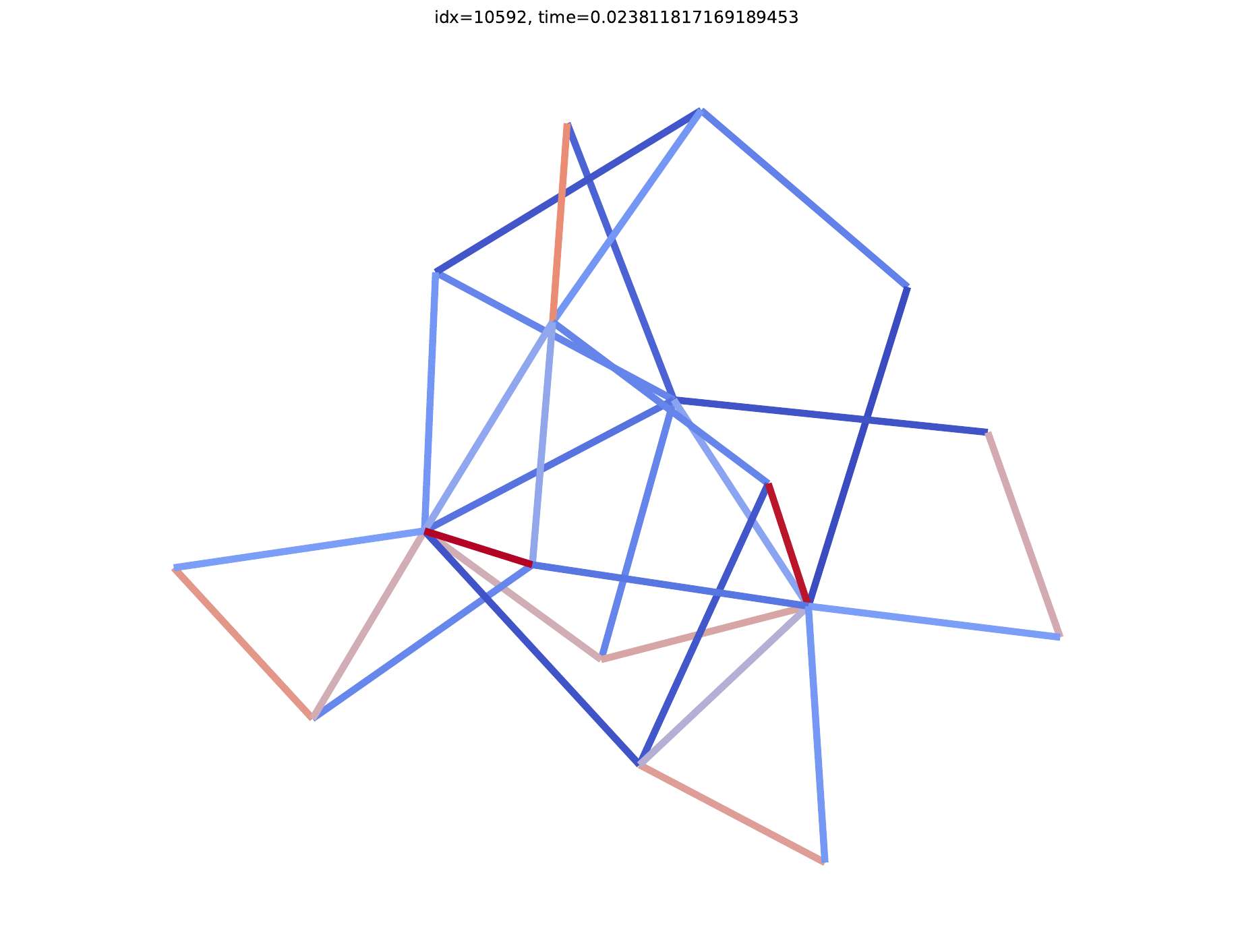} &
\imgcell{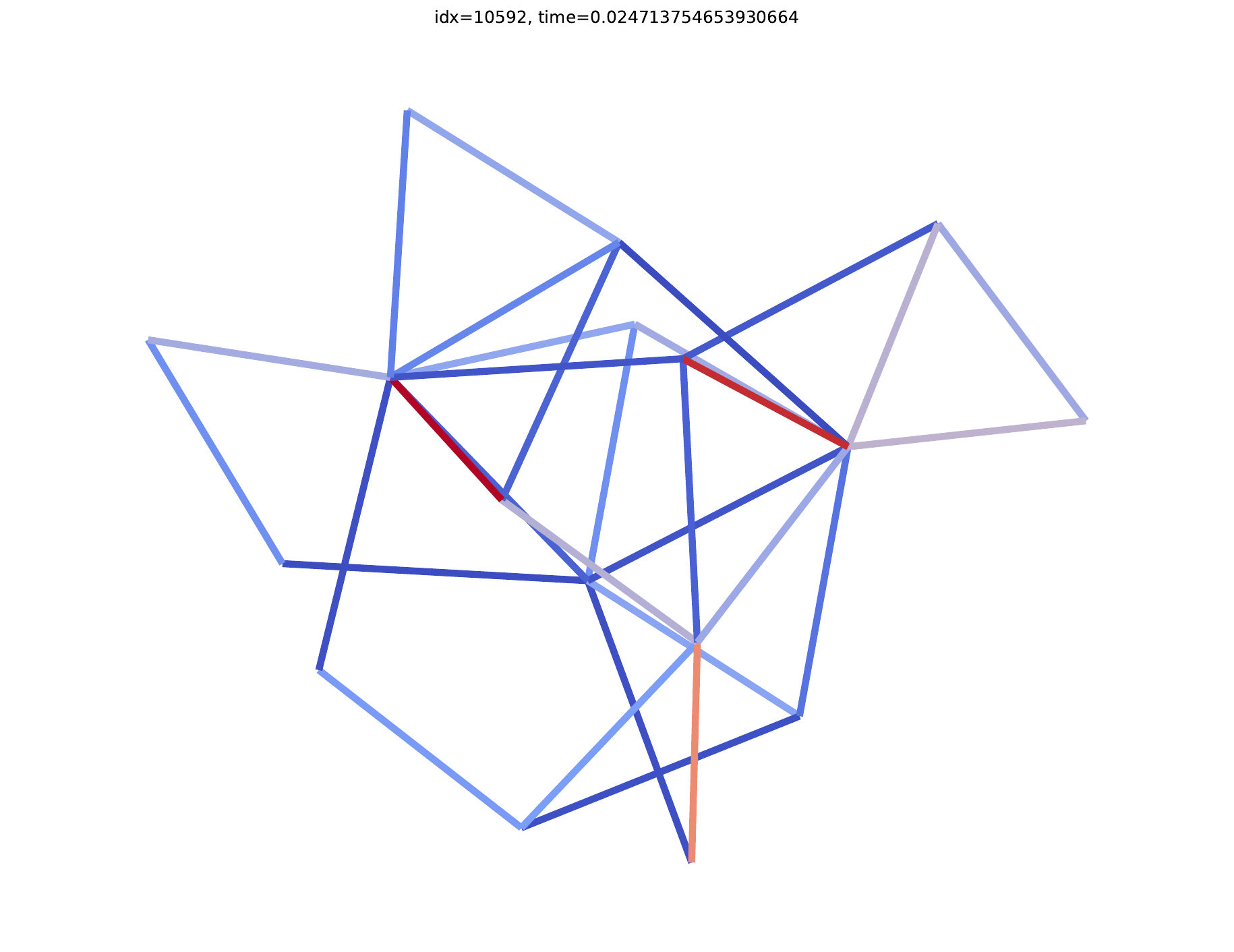} \\

&
t = 0.00s &
t = 0.26s &
t = 0.03s &
t = 0.04s &
t = 100.42s &
t = 0.02s &
t = 0.02s &
t = 0.02s &
t = 0.02s &
t = 0.02s &
t = 0.02s &
t = 0.02s \\

\makecell{\bfseries grafo10635.95\\N = 40\\M = 56} &
\imgcell{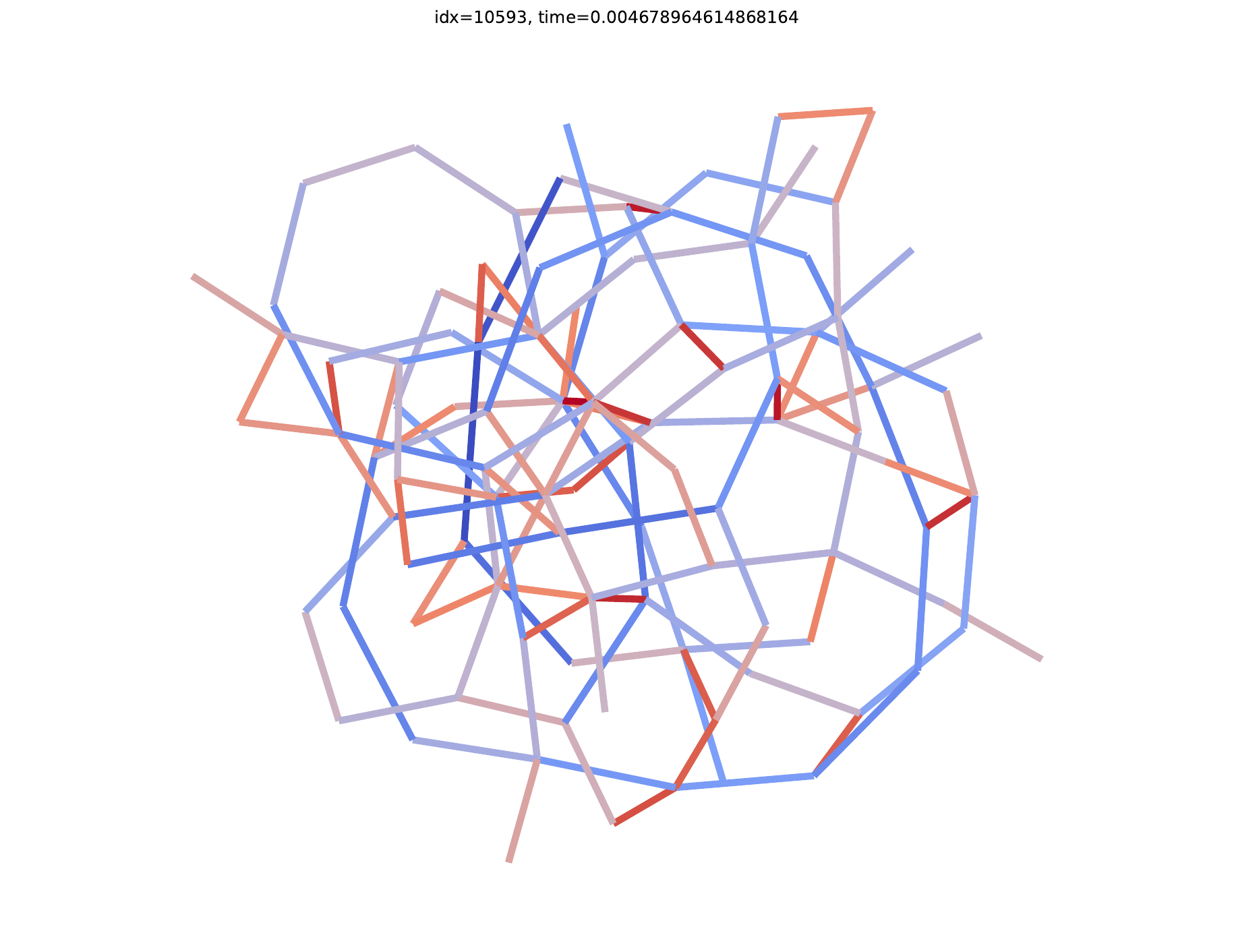} &
\imgcell{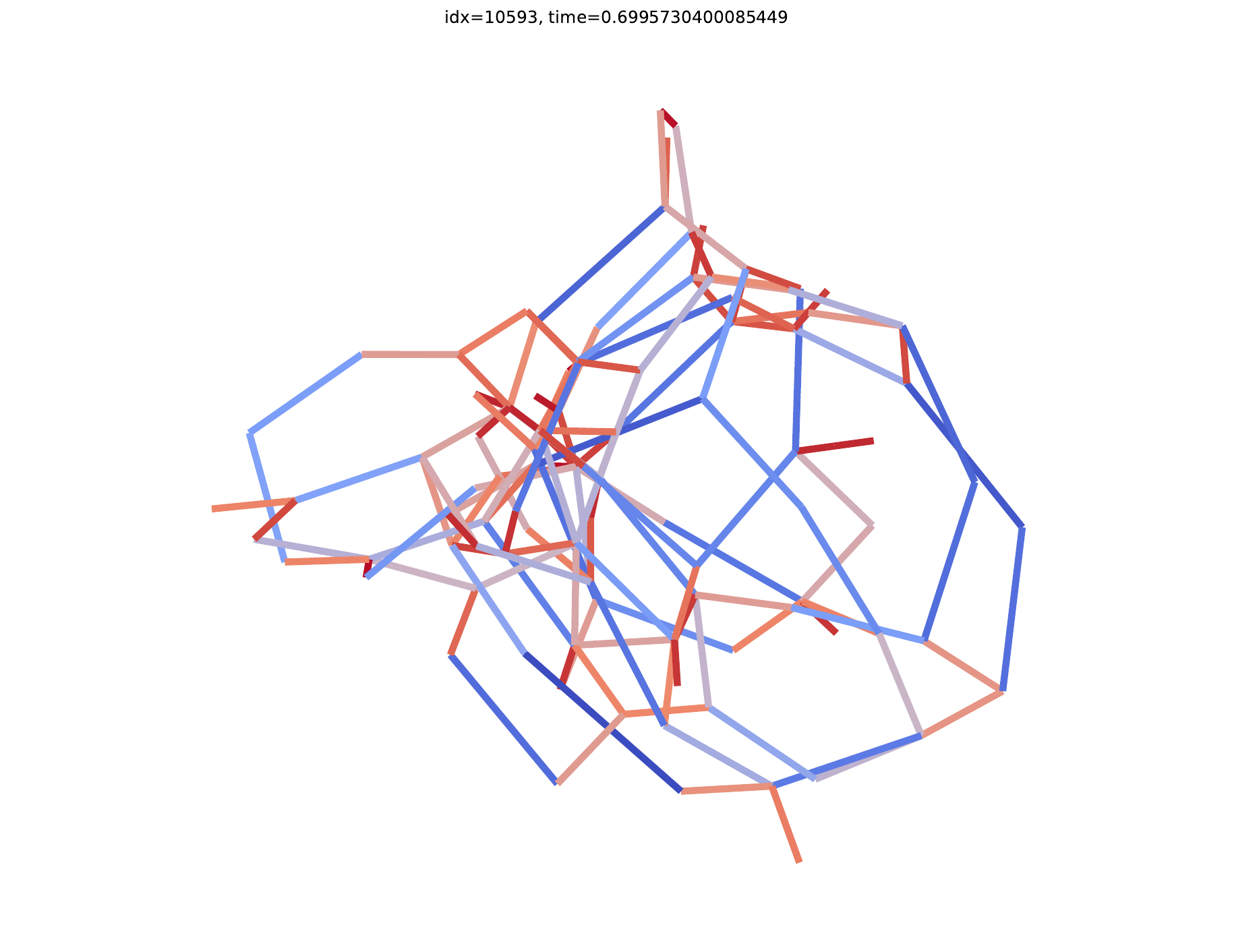} &
\imgcell{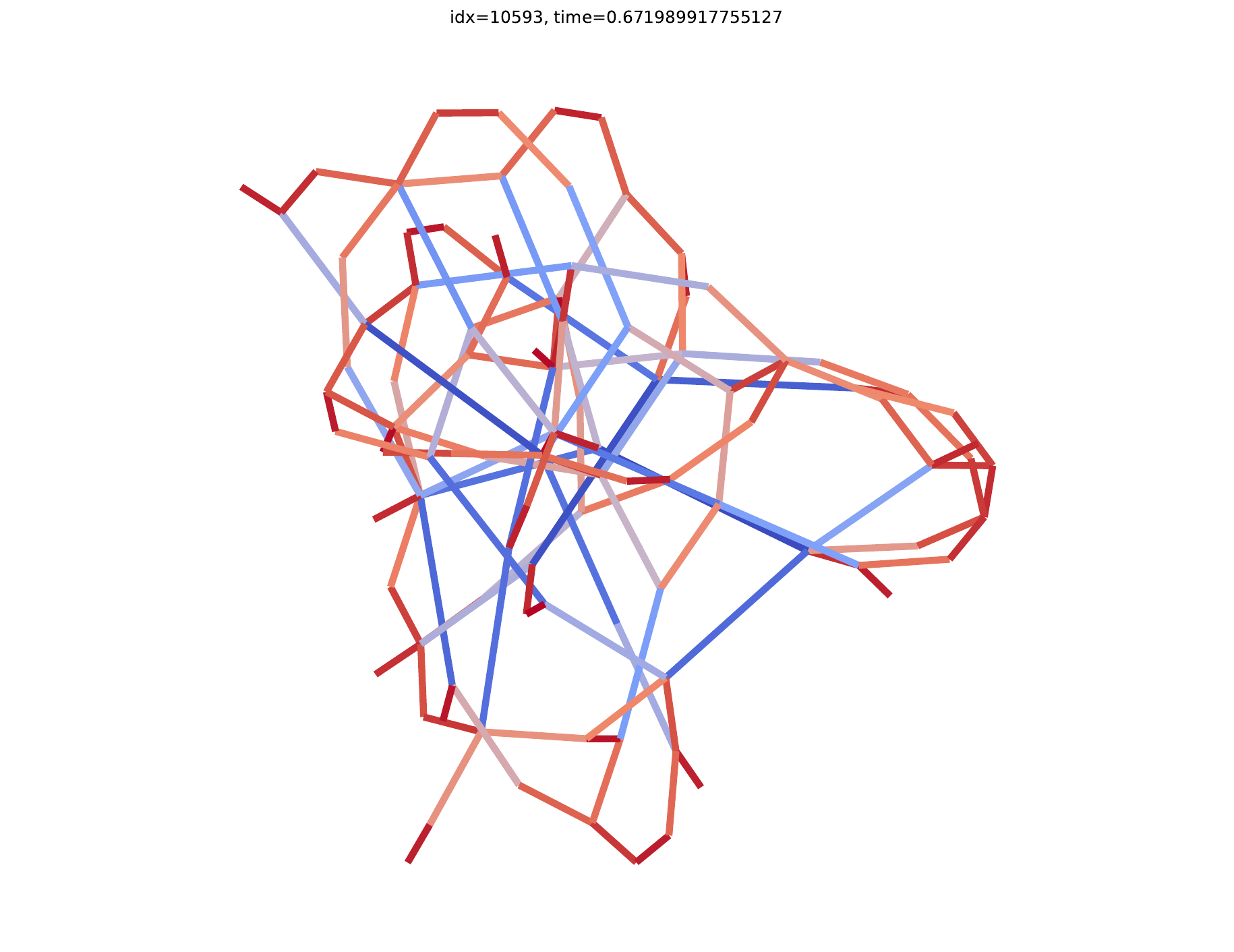} &
\imgcell{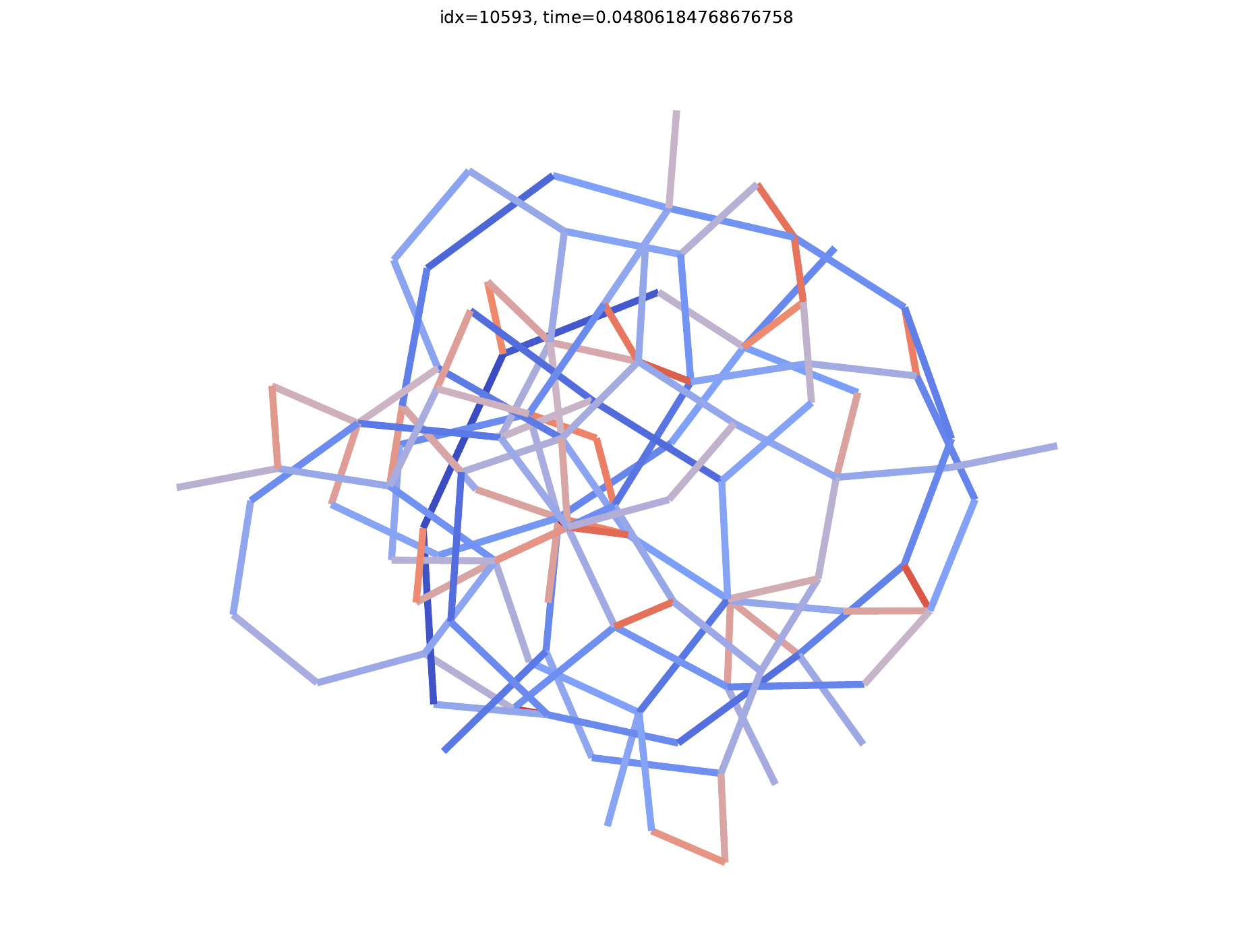} &
\imgcell{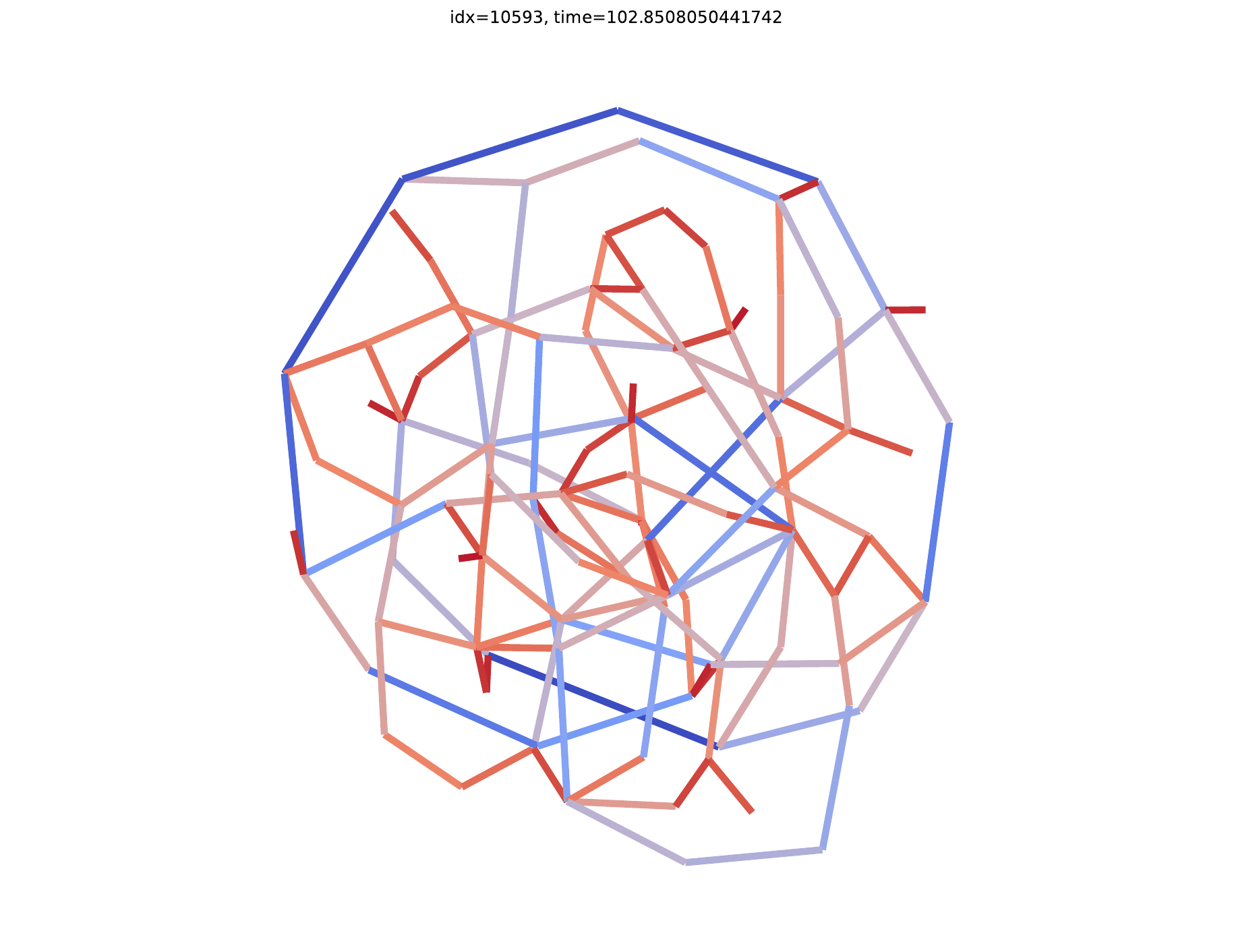} &
\imgcell{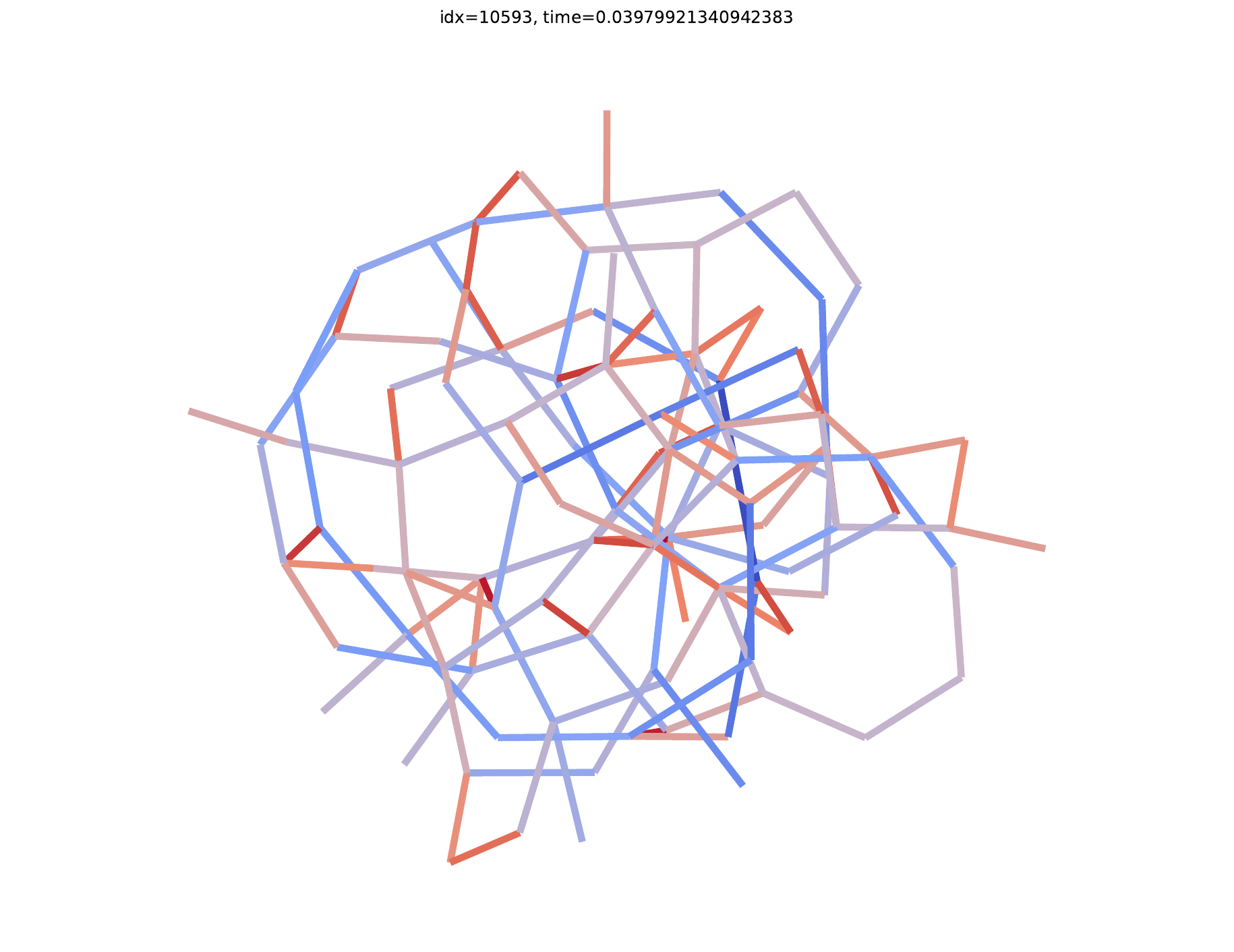} &
\imgcell{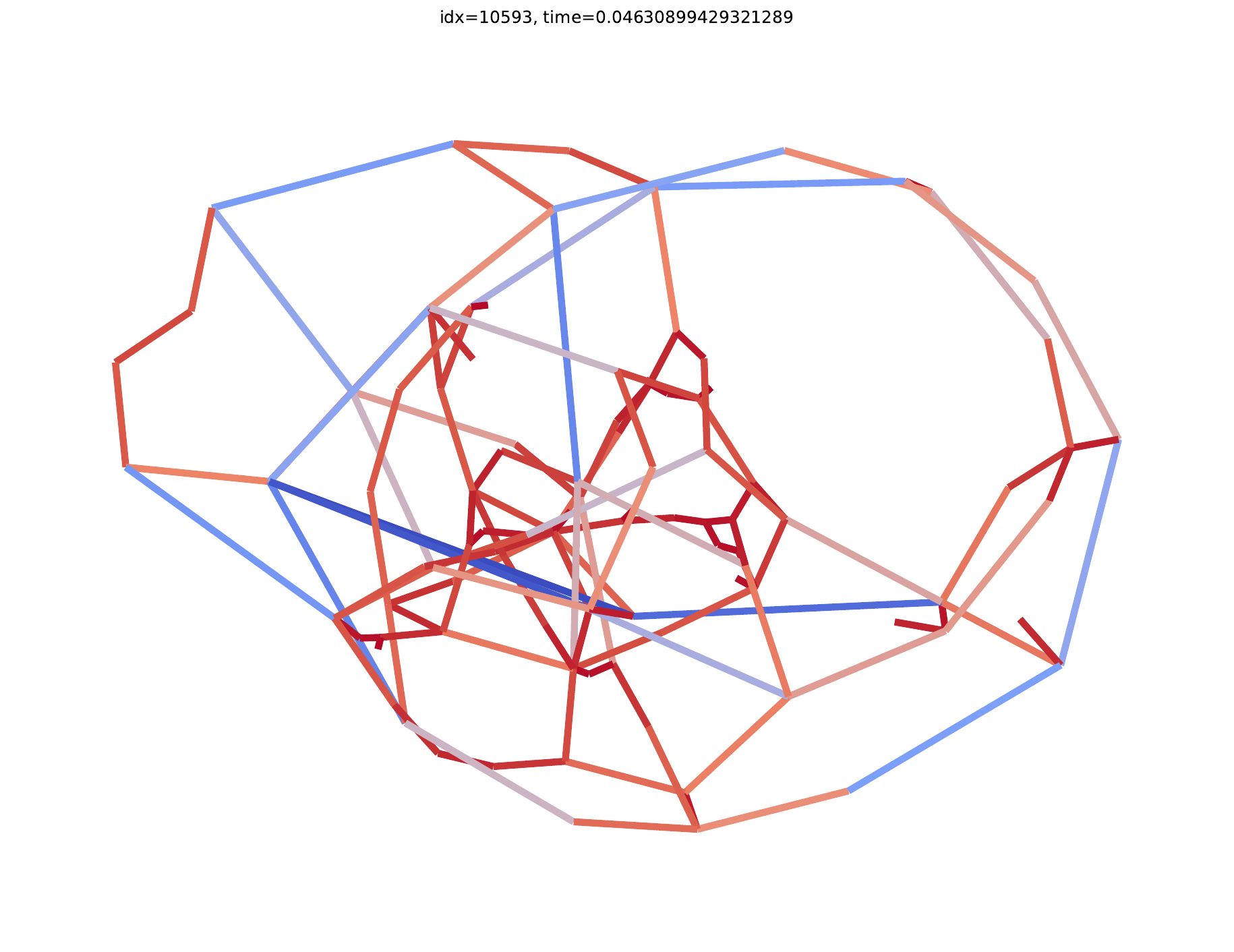} &
\imgcell{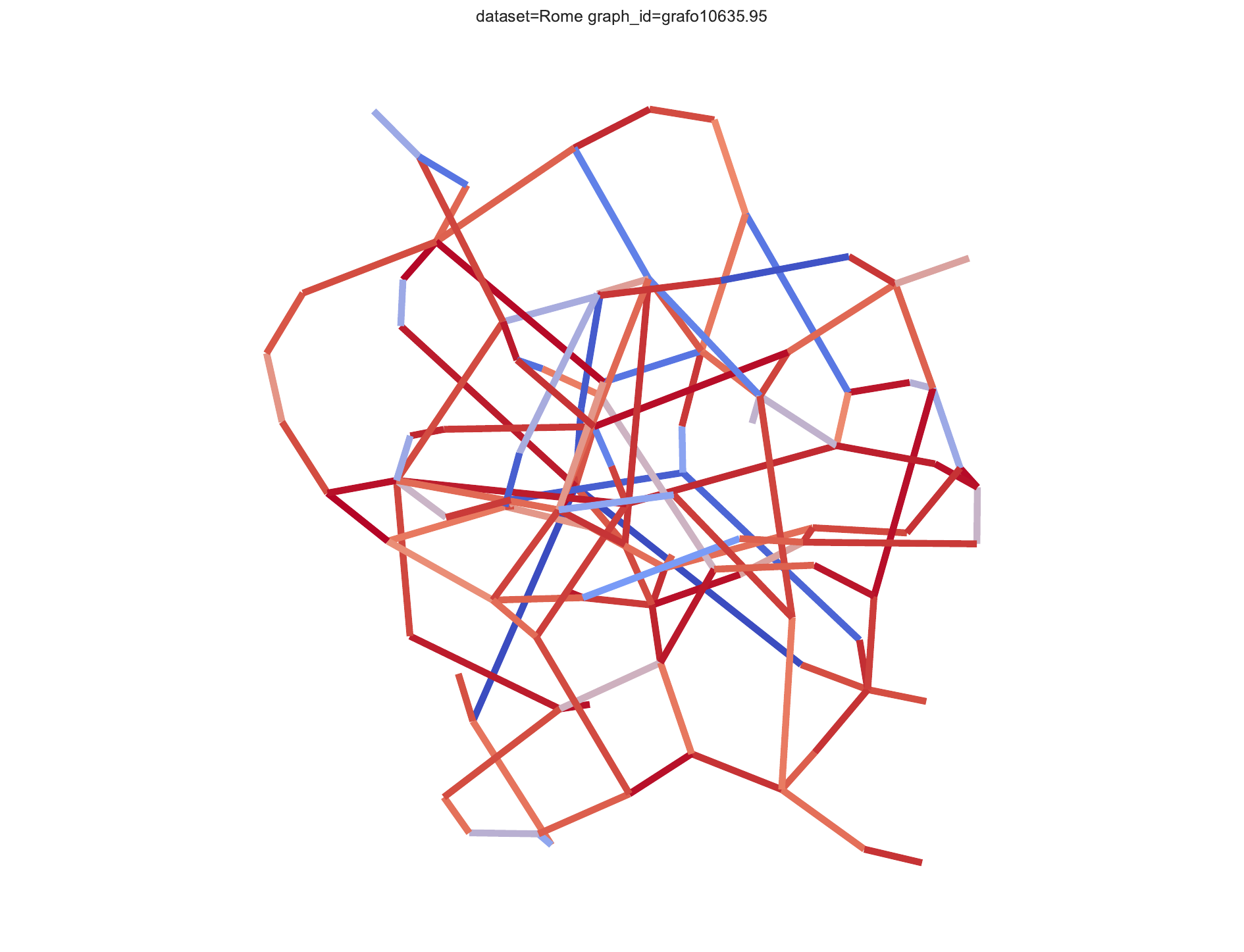} &
\imgcell{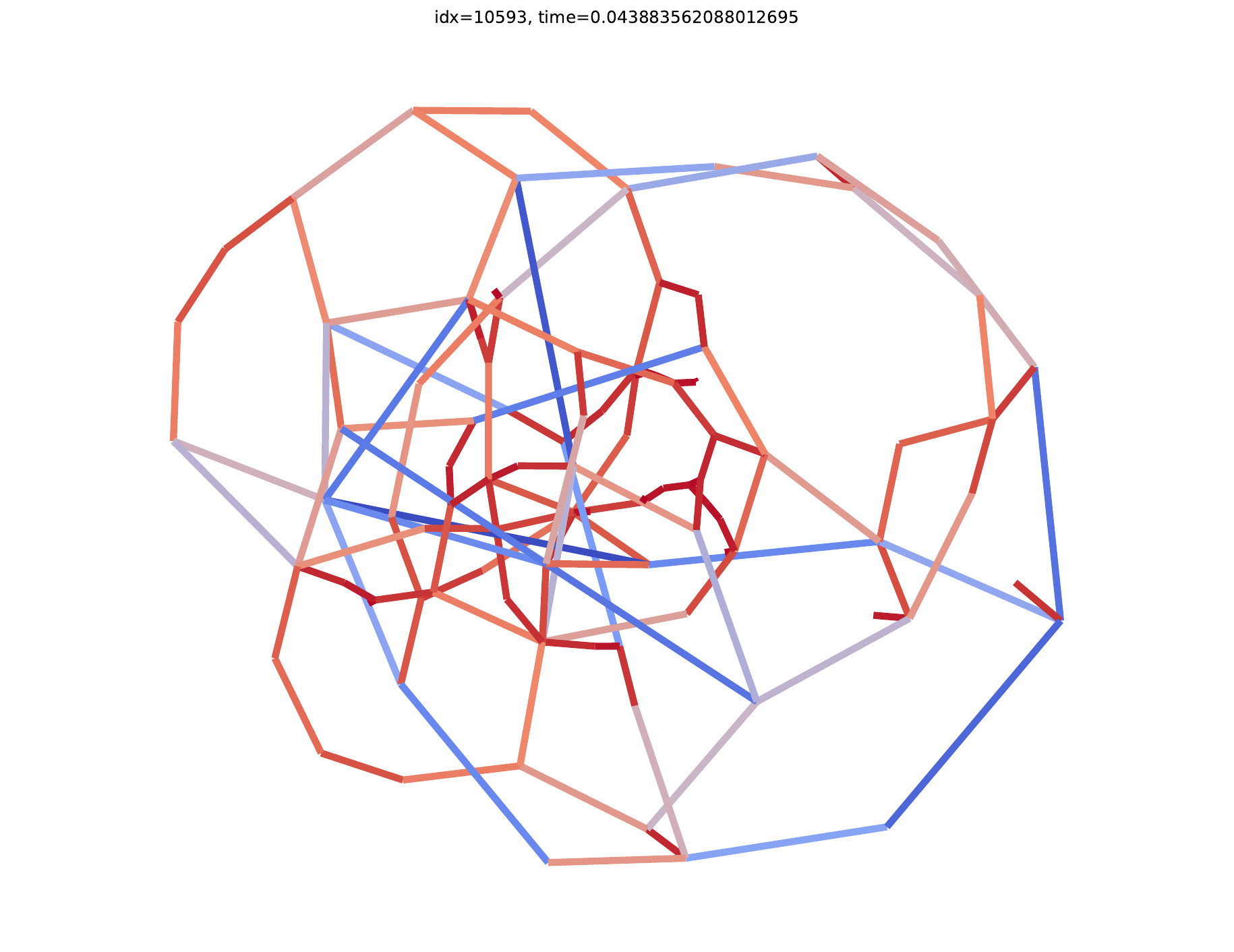} &
\imgcell{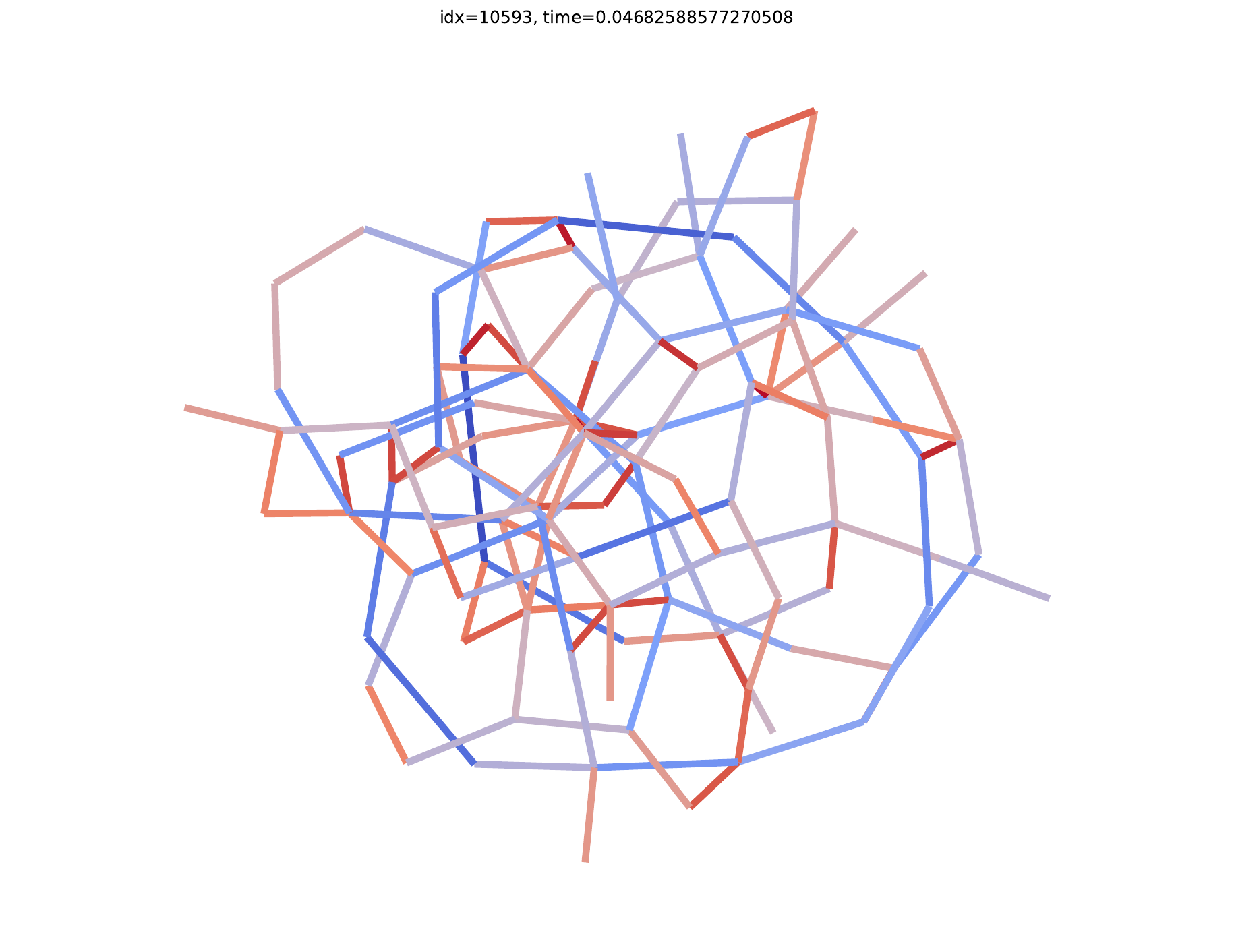} &
\imgcell{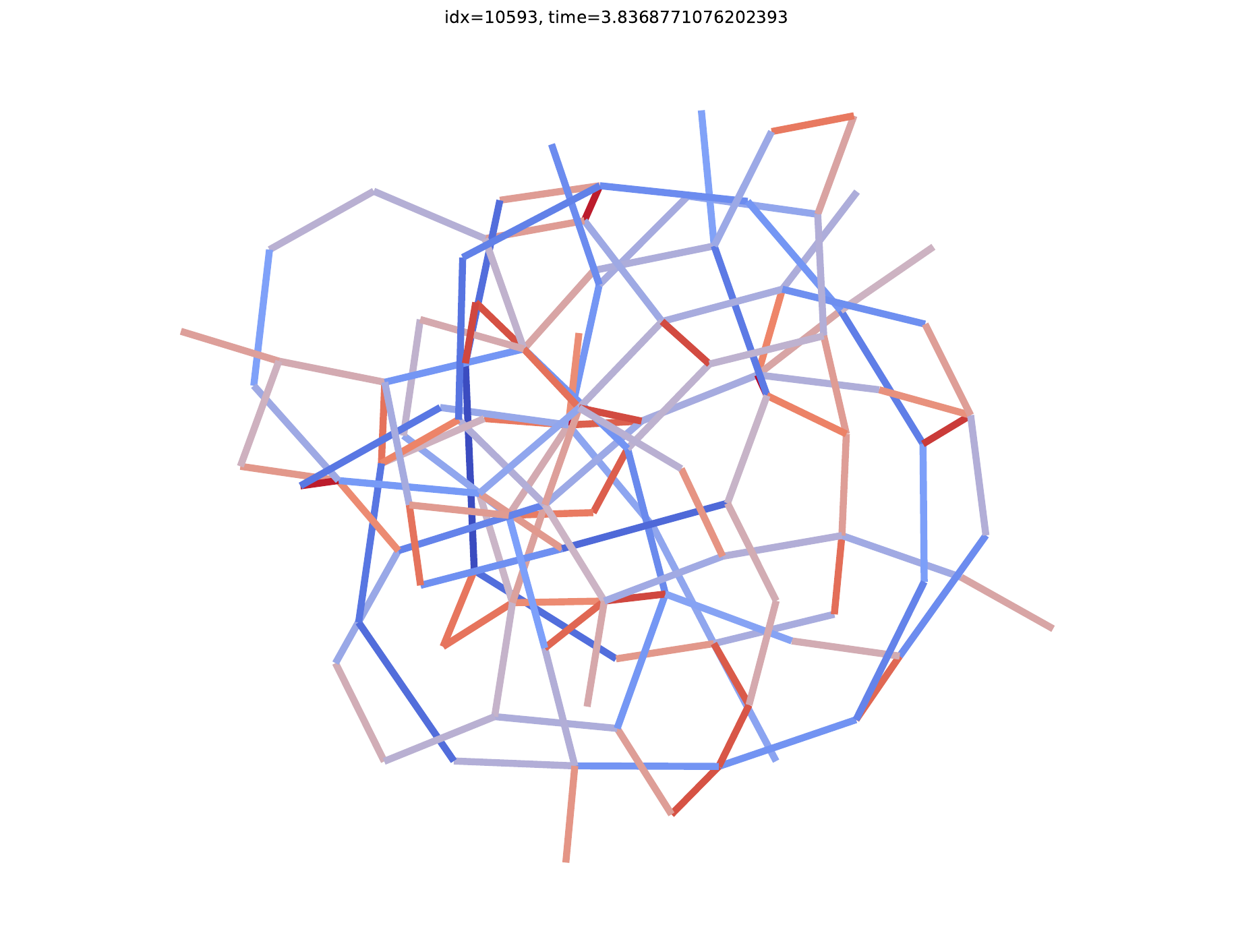} &
\imgcell{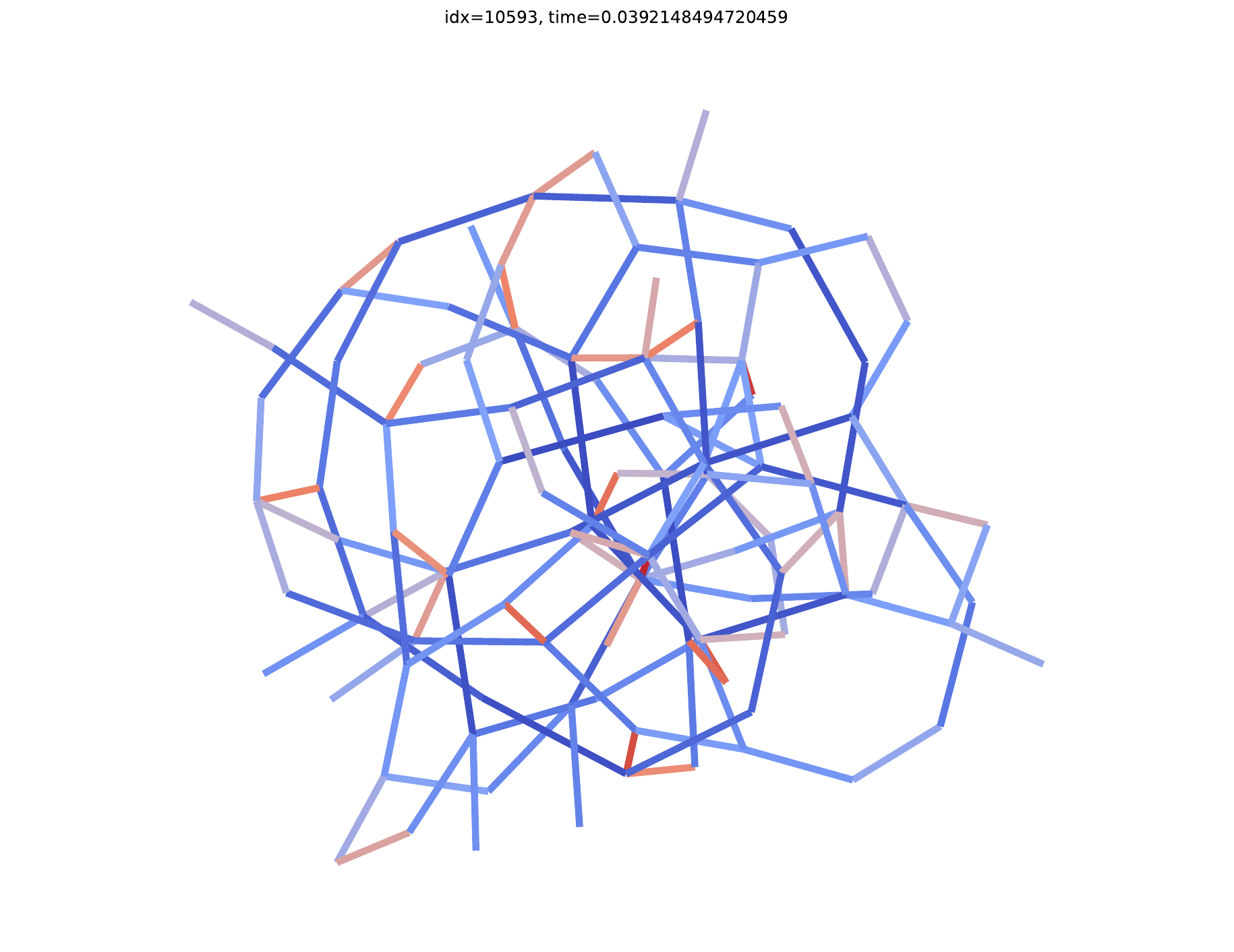} \\

&
t = 0.00s &
t = 0.70s &
t = 0.67s &
t = 0.05s &
t = 102.85s &
t = 0.04s &
t = 0.05s &
t = 0.04s &
t = 0.04s &
t = 0.05s &
t = 0.04s &
t = 0.04s \\

\makecell{\bfseries grafo10464.96\\N = 77\\M = 92} &
\imgcell{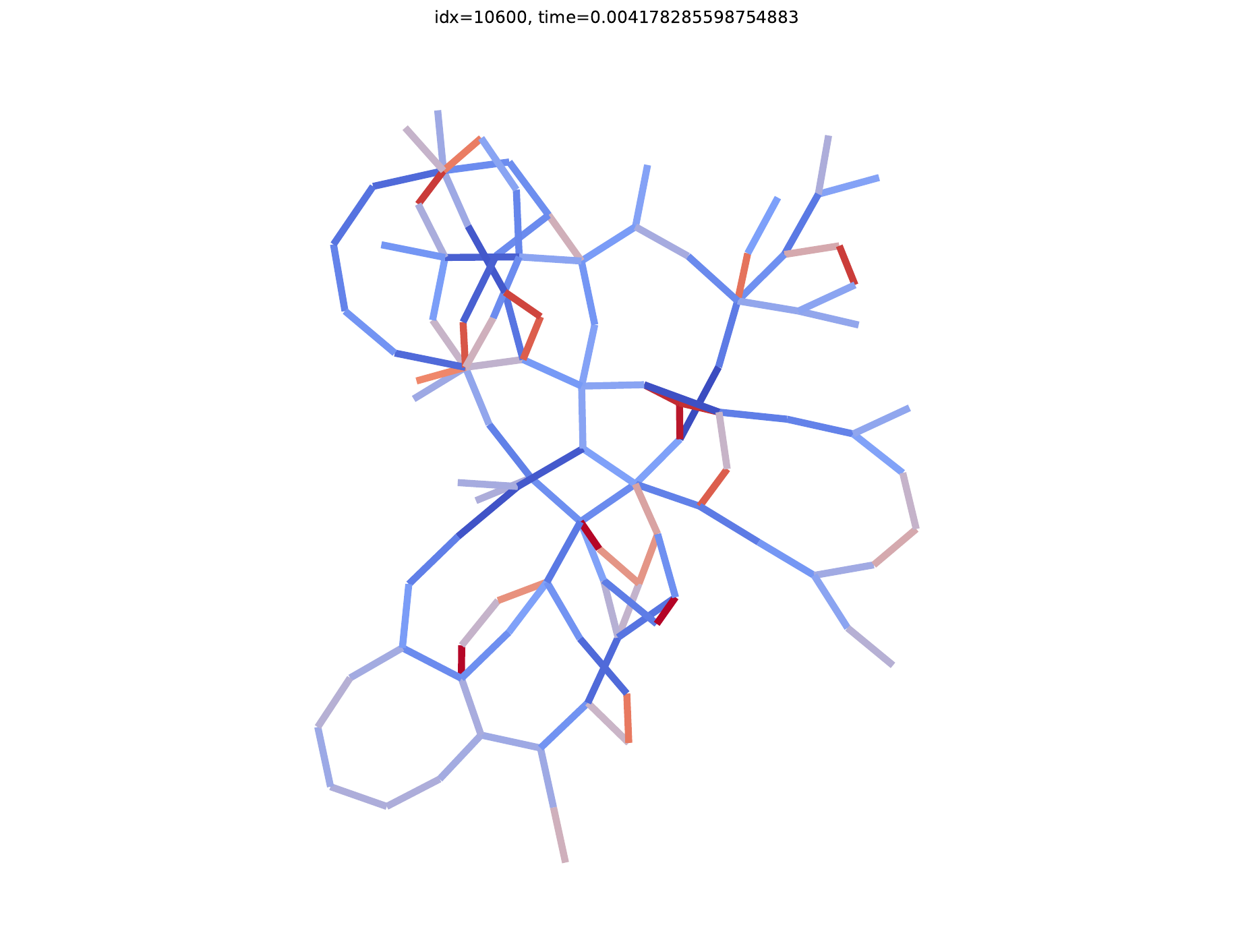} &
\imgcell{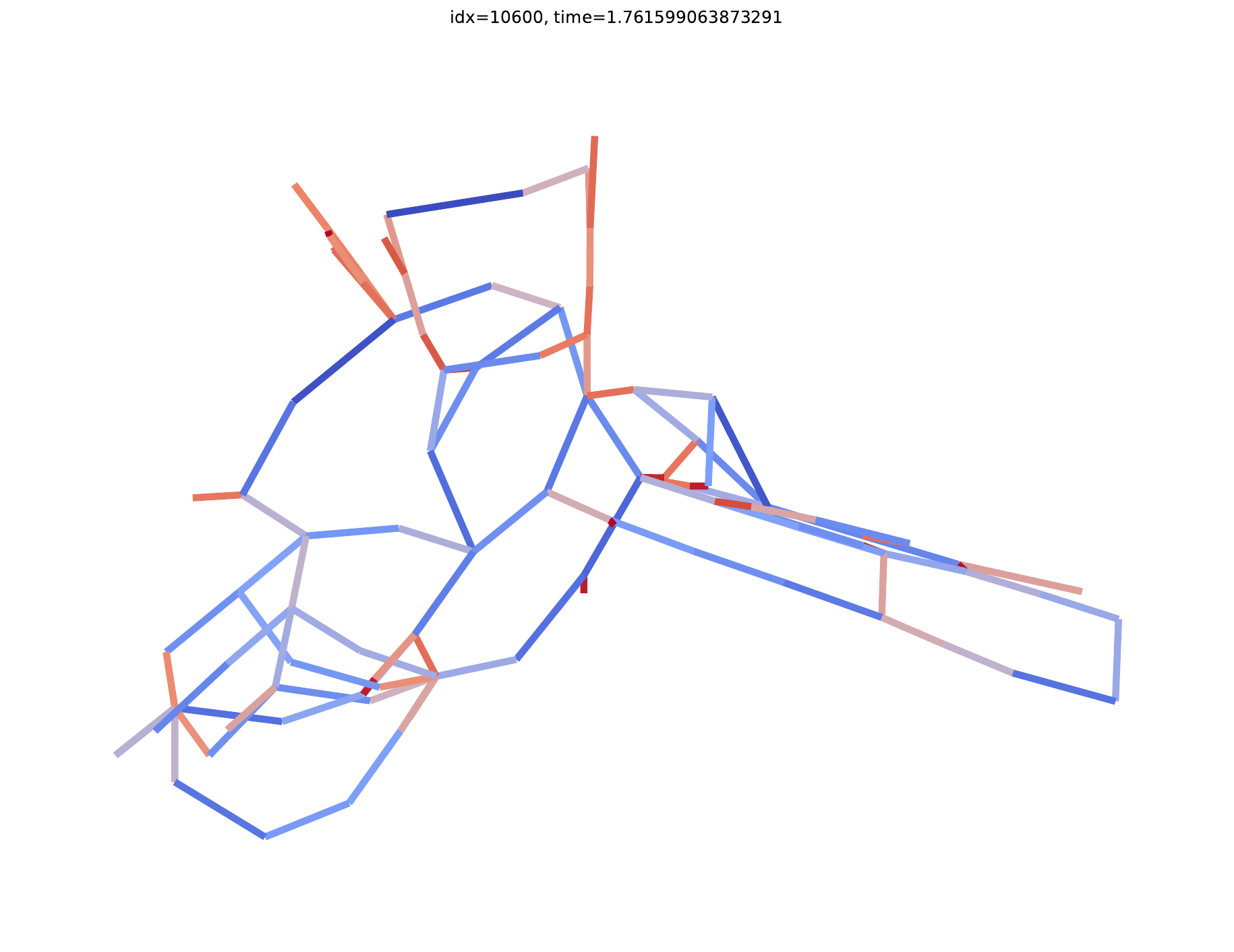} &
\imgcell{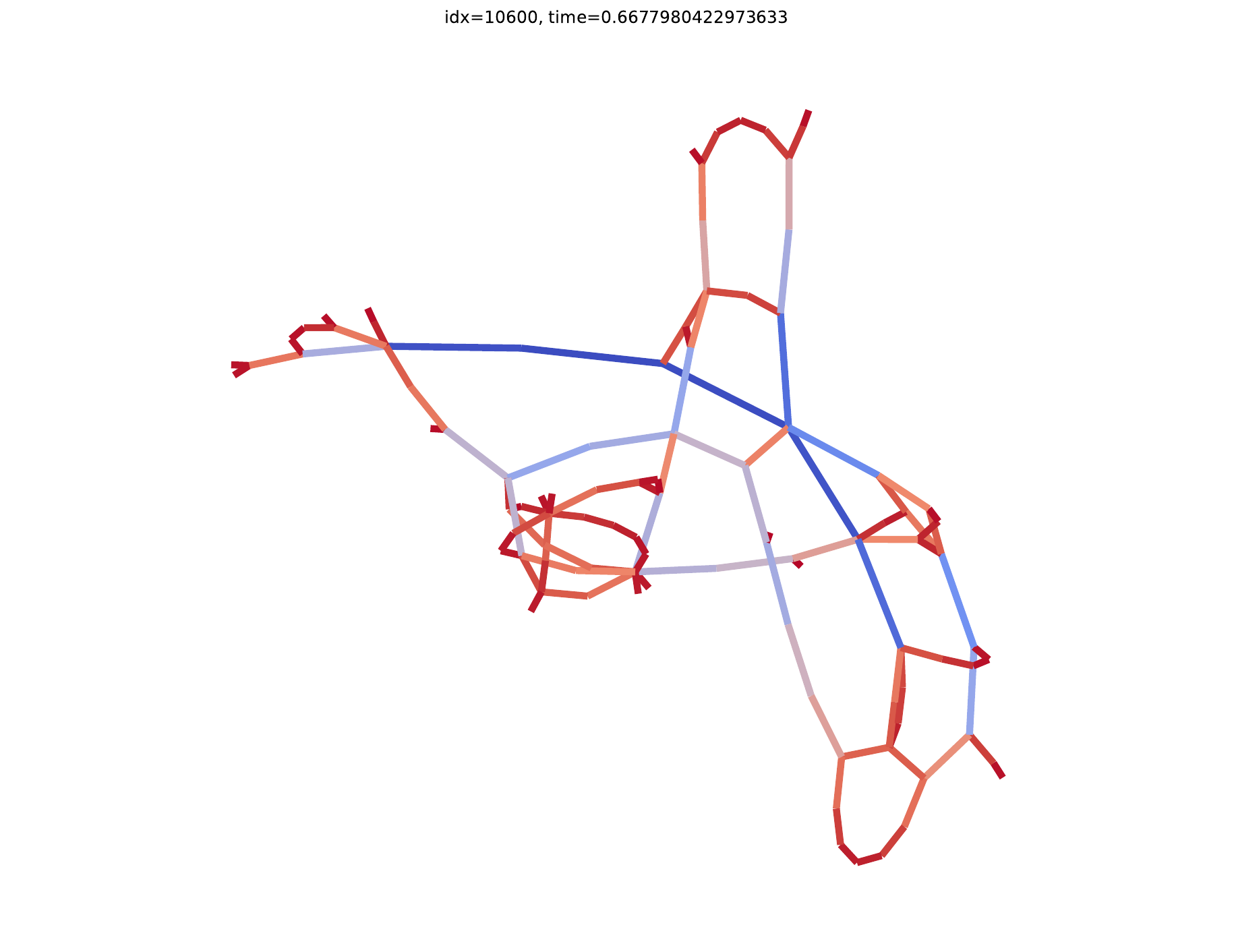} &
\imgcell{figures/rome_graphs/10600_deepgd.pdf} &
\imgcell{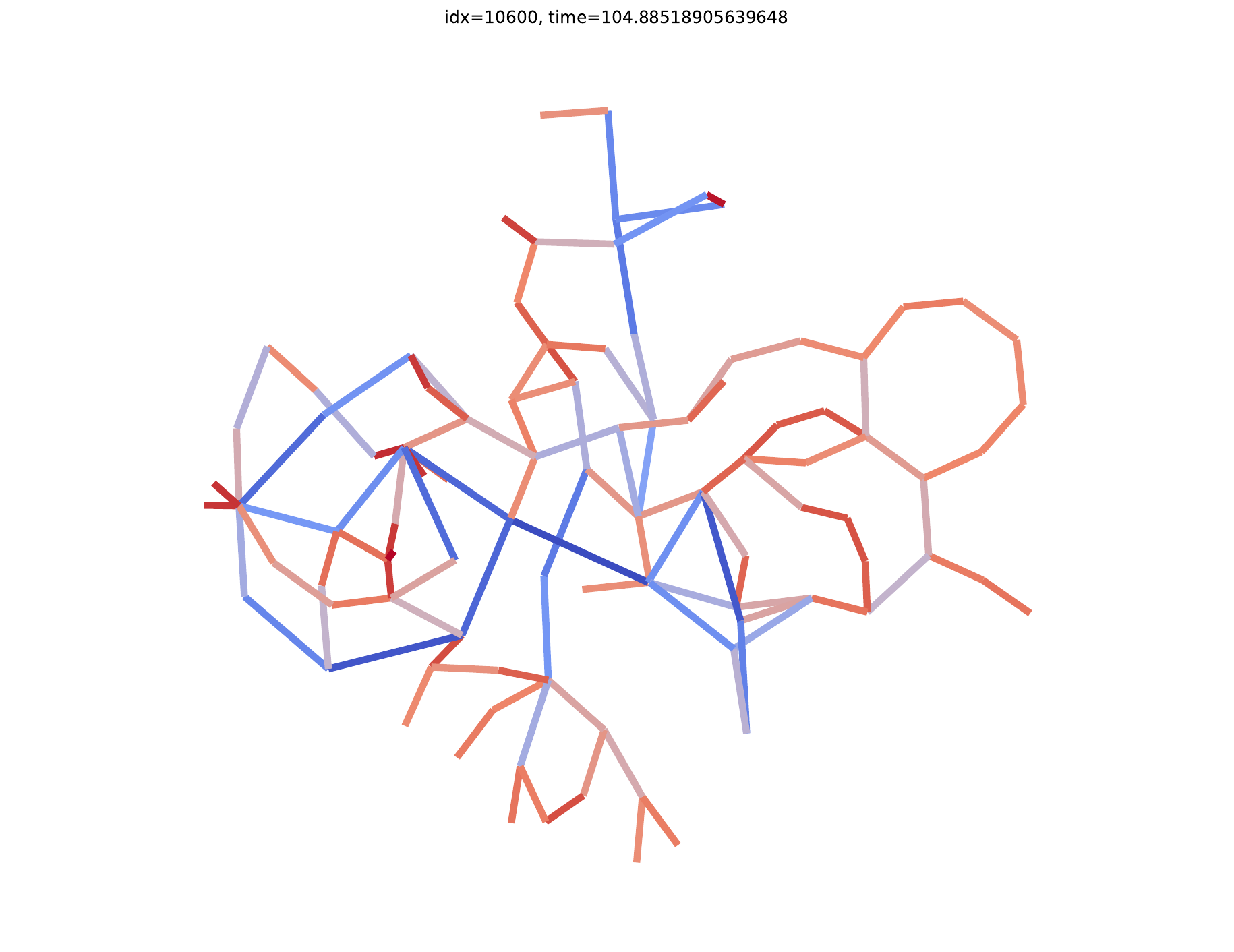} &
\imgcell{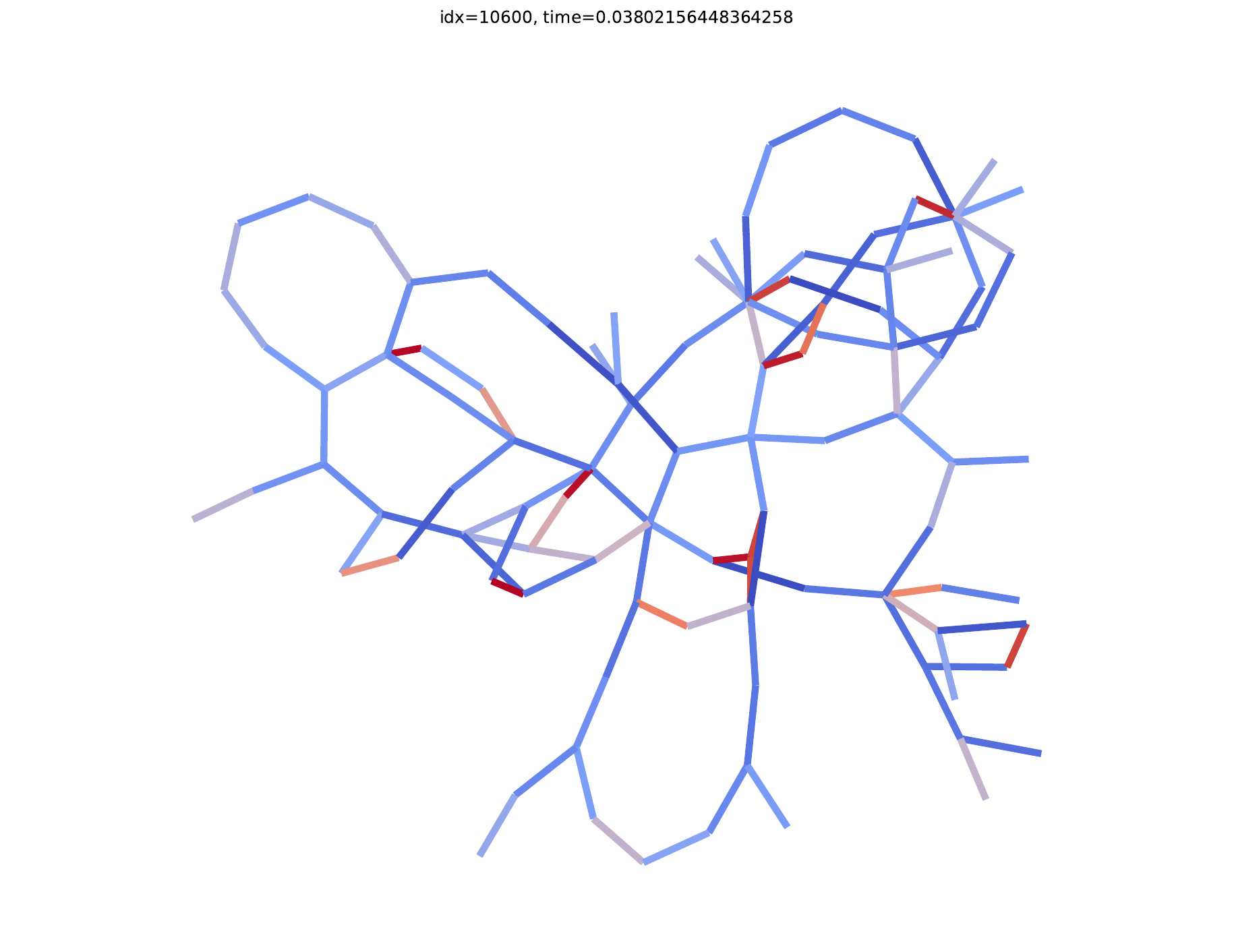} &
\imgcell{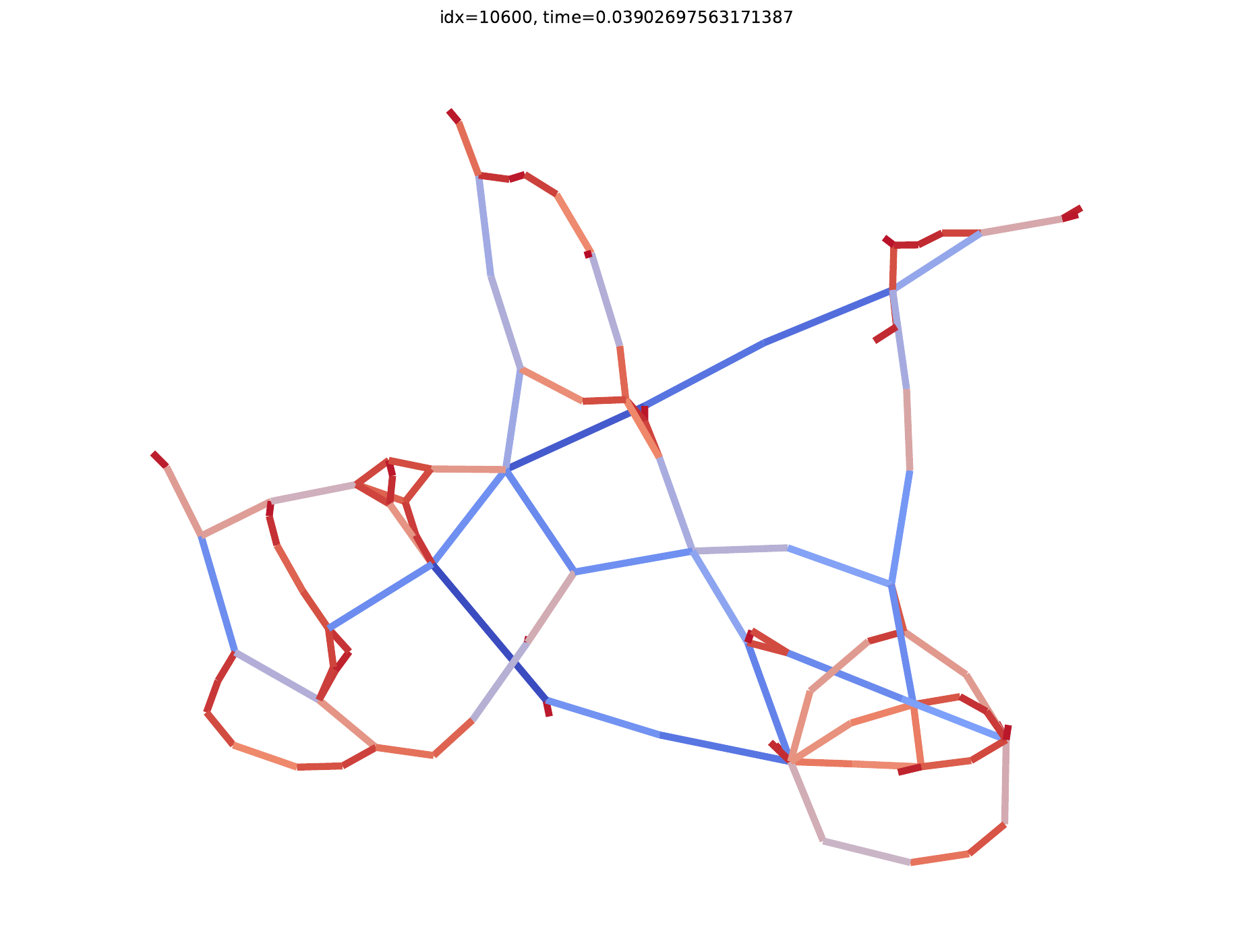} &
\imgcell{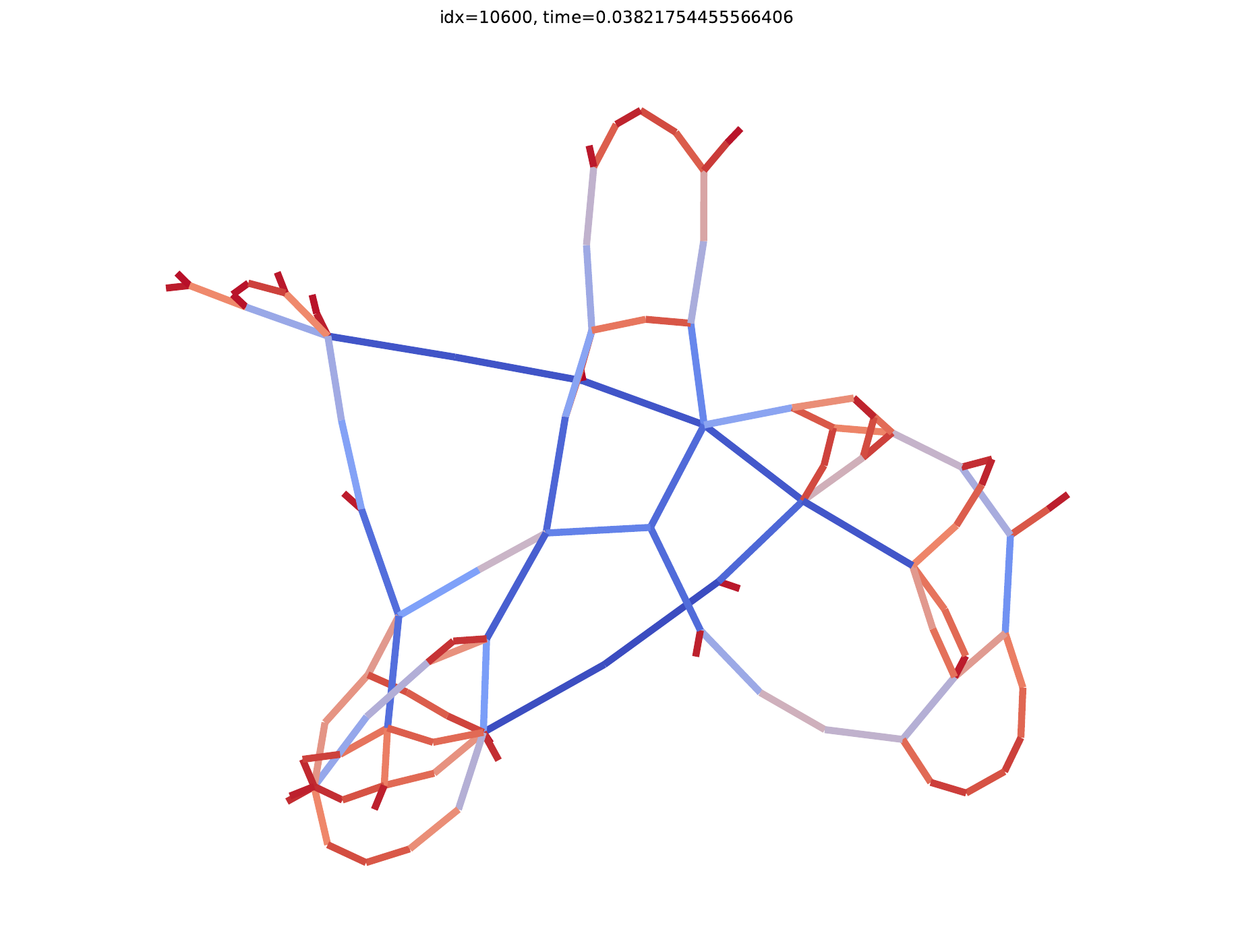} &
\imgcell{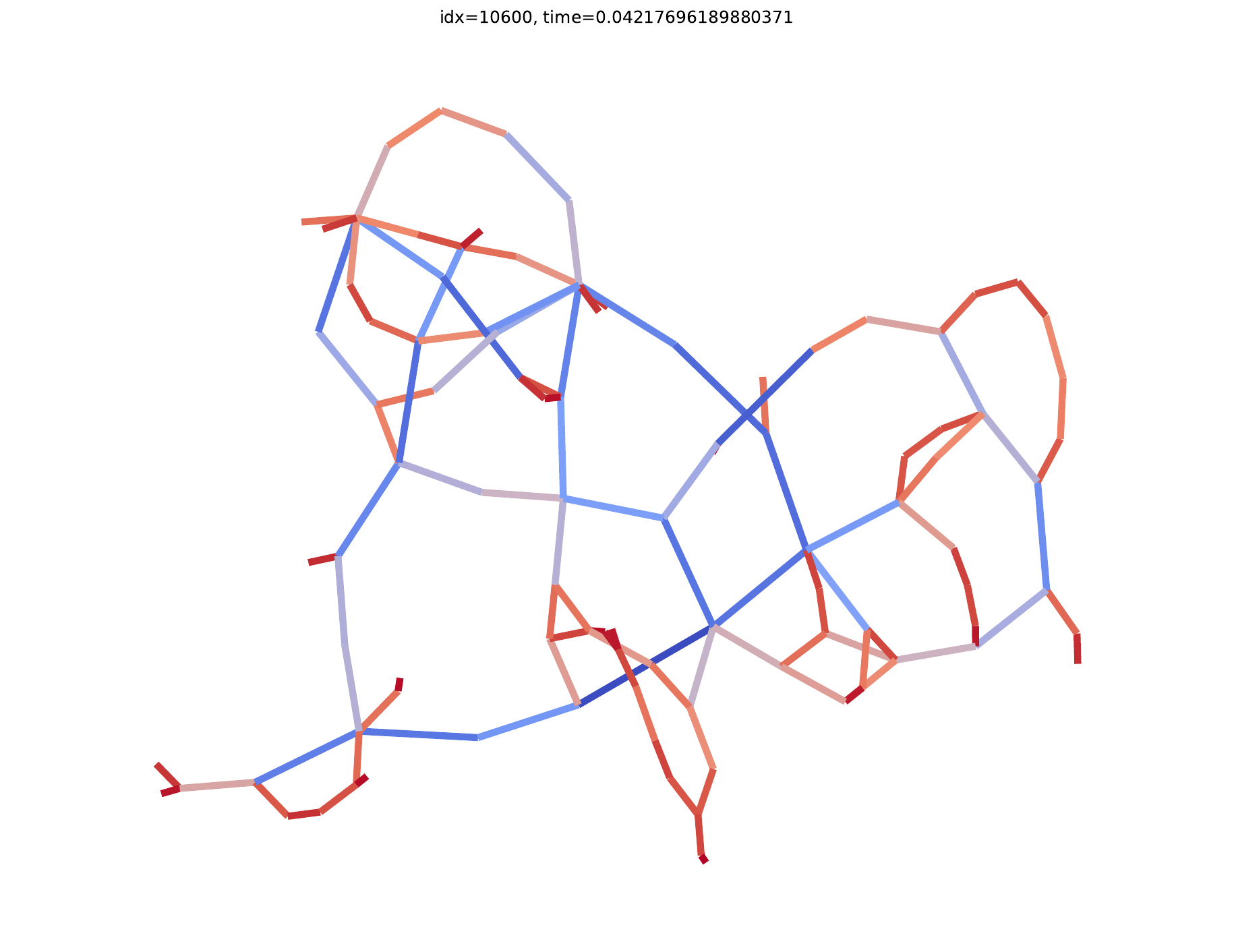} &
\imgcell{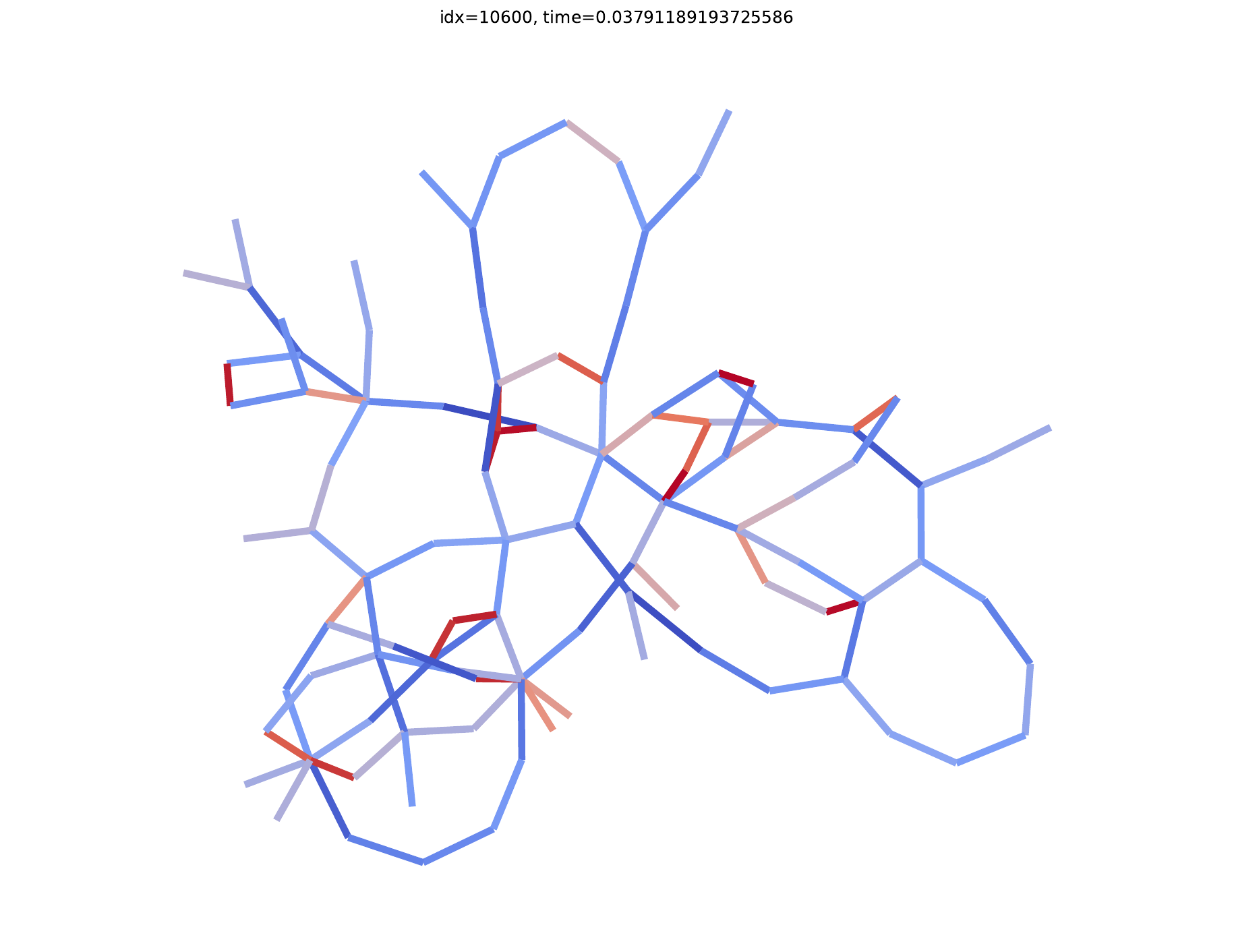} &
\imgcell{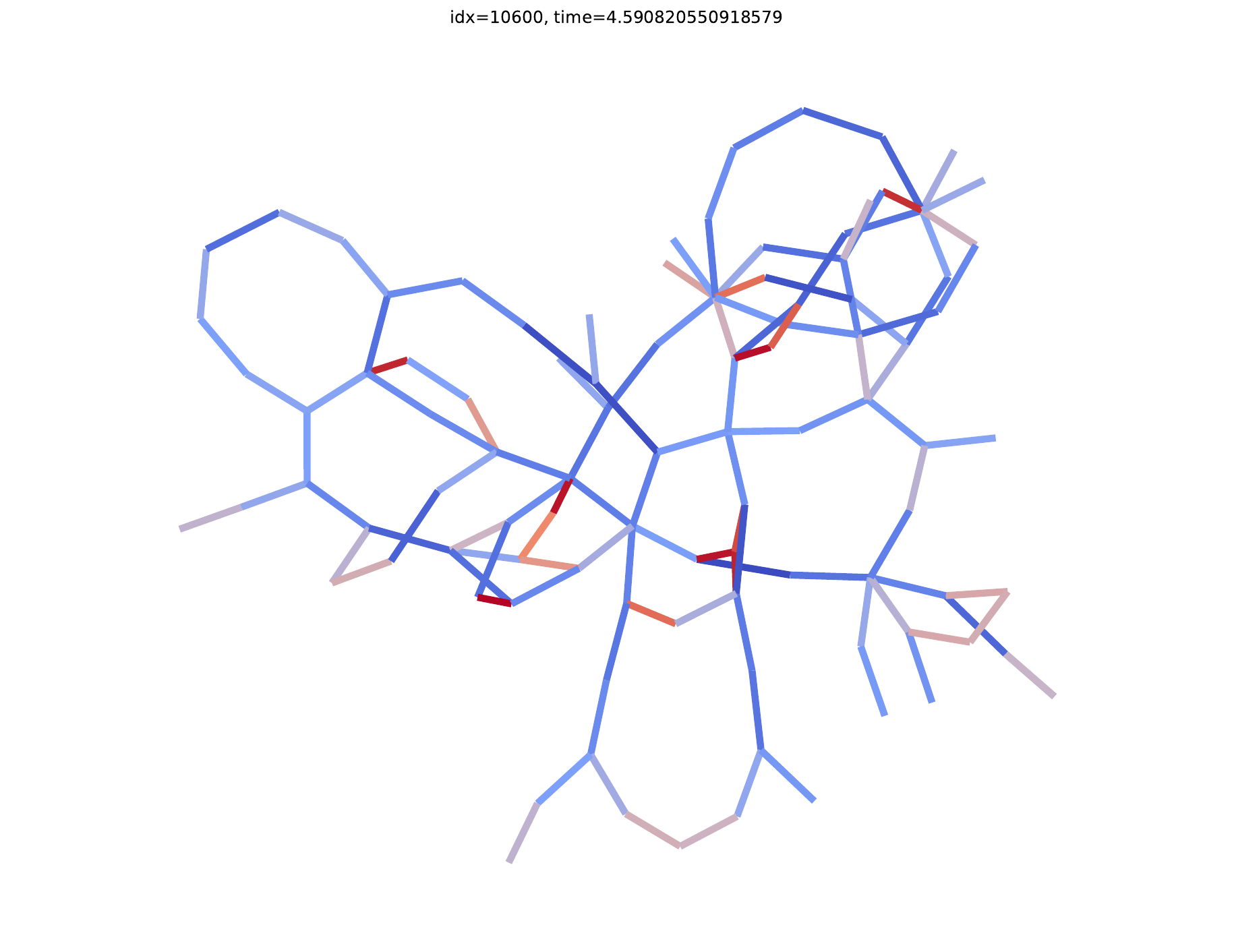} &
\imgcell{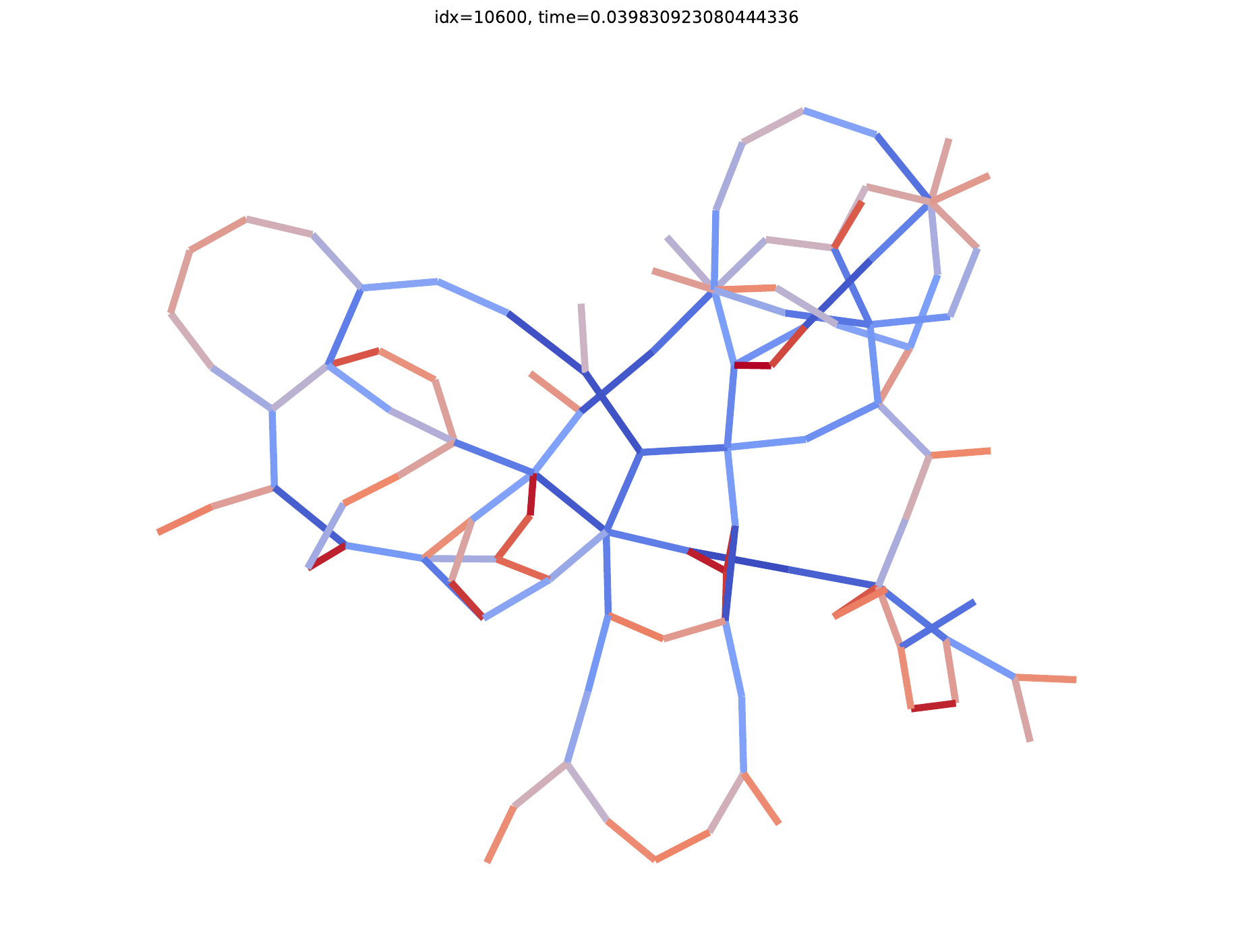} \\

&
t = 0.00s &
t = 1.76s &
t = 0.67s &
t = 0.05s &
t = 104.89s &
t = 0.04s &
t = 0.04s &
t = 0.04s &
t = 0.04s &
t = 0.04s &
t = 0.05s &
t = 0.04s \\

\makecell{\bfseries grafo672.39\\N = 23\\M = 28} &
\imgcell{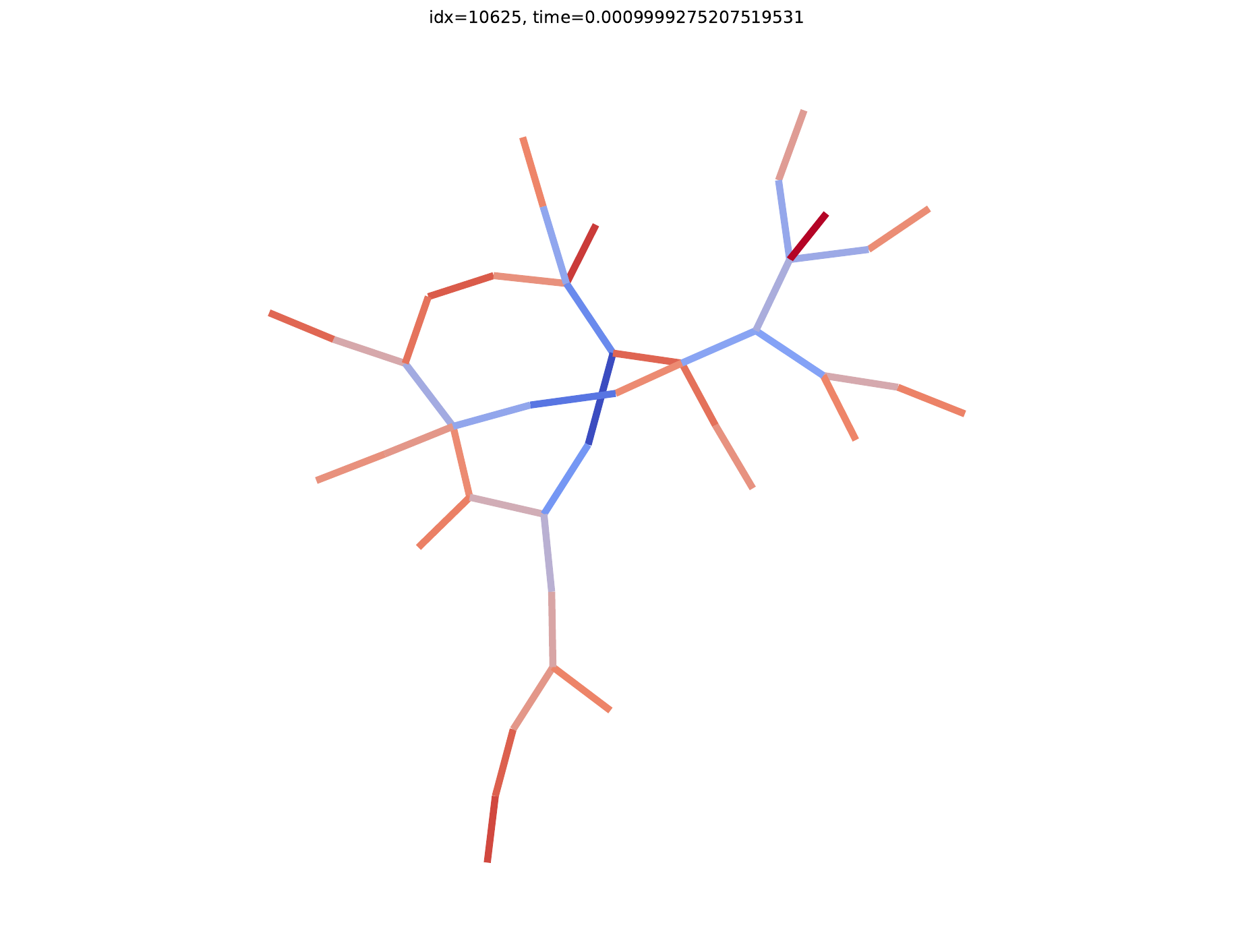} &
\imgcell{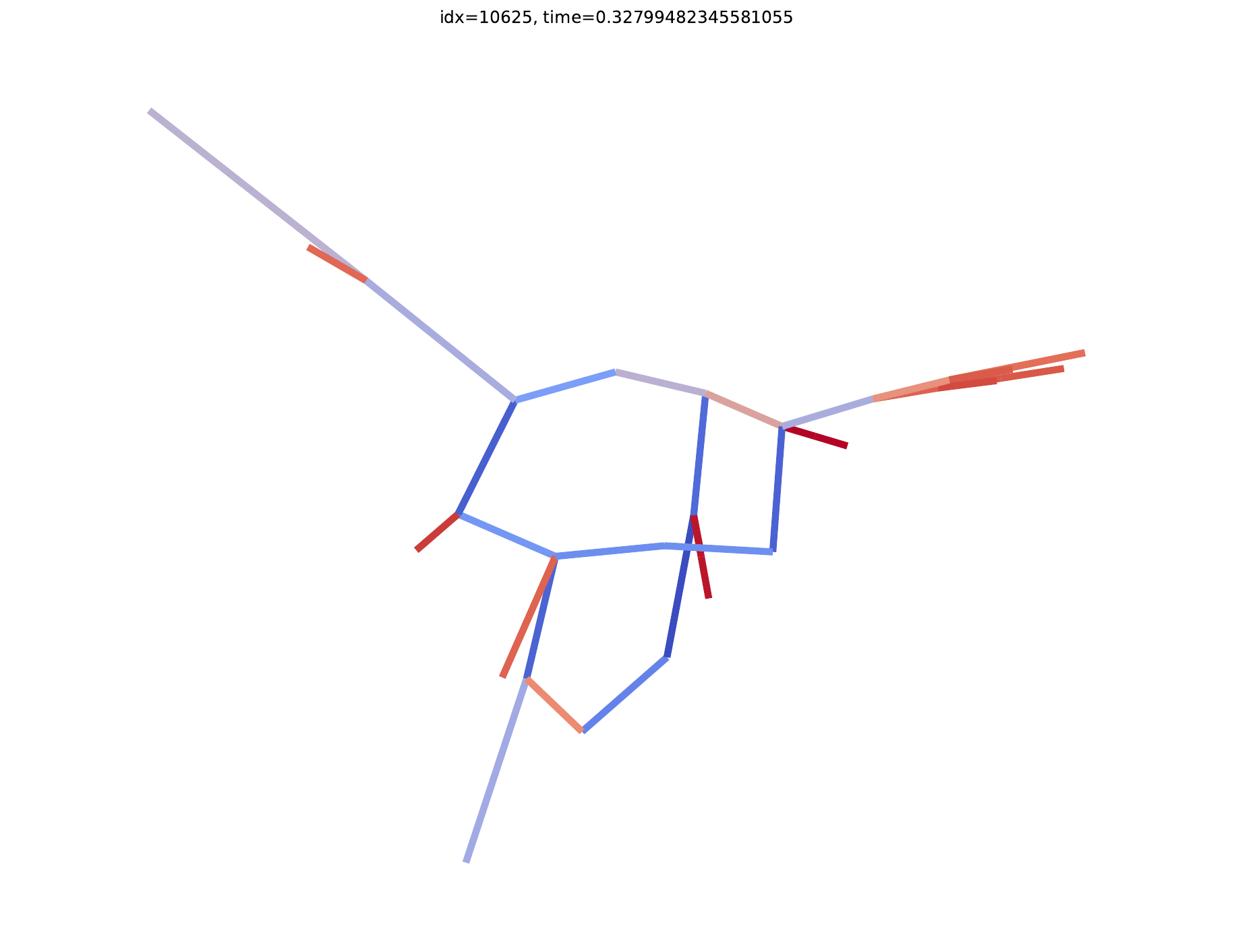} &
\imgcell{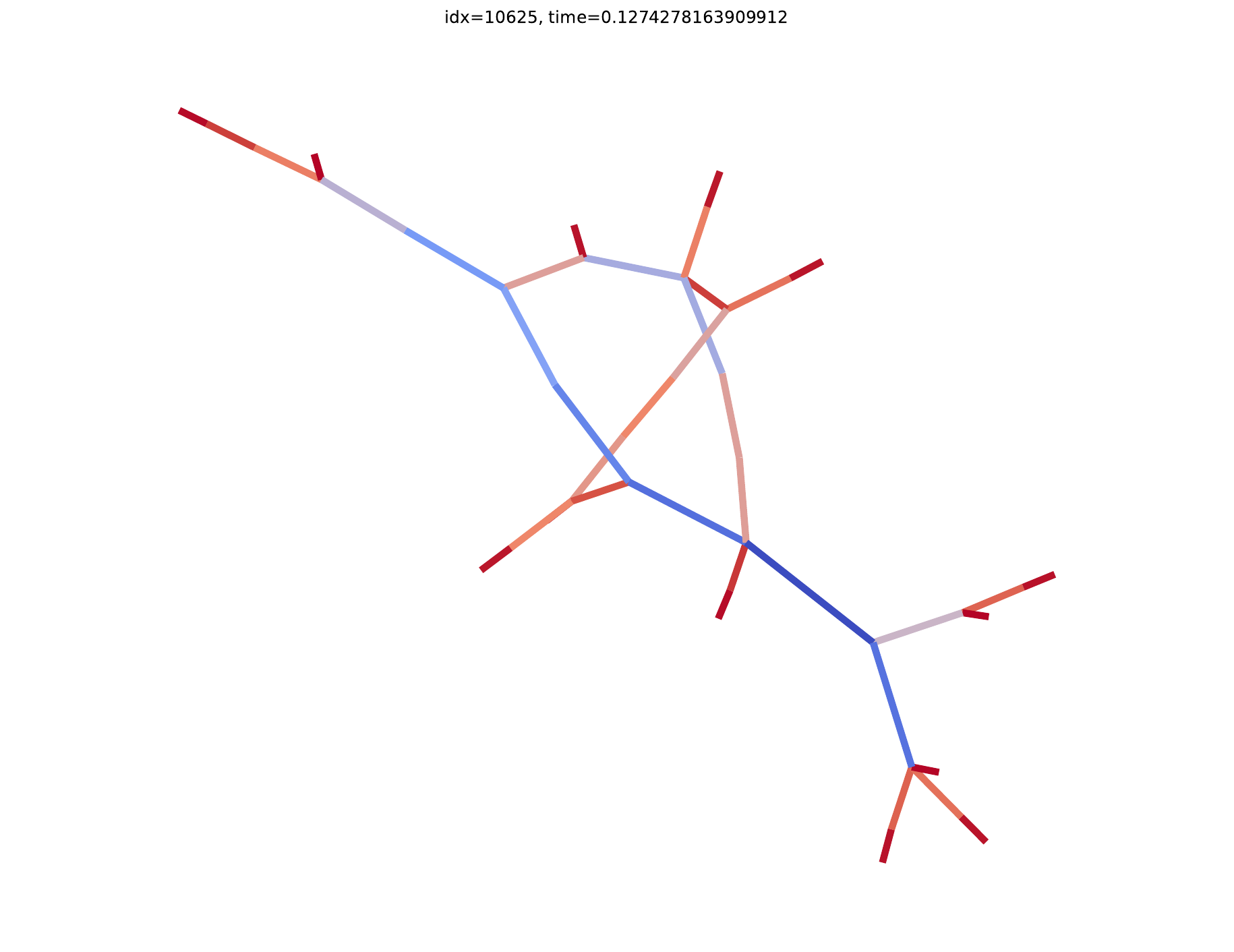} &
\imgcell{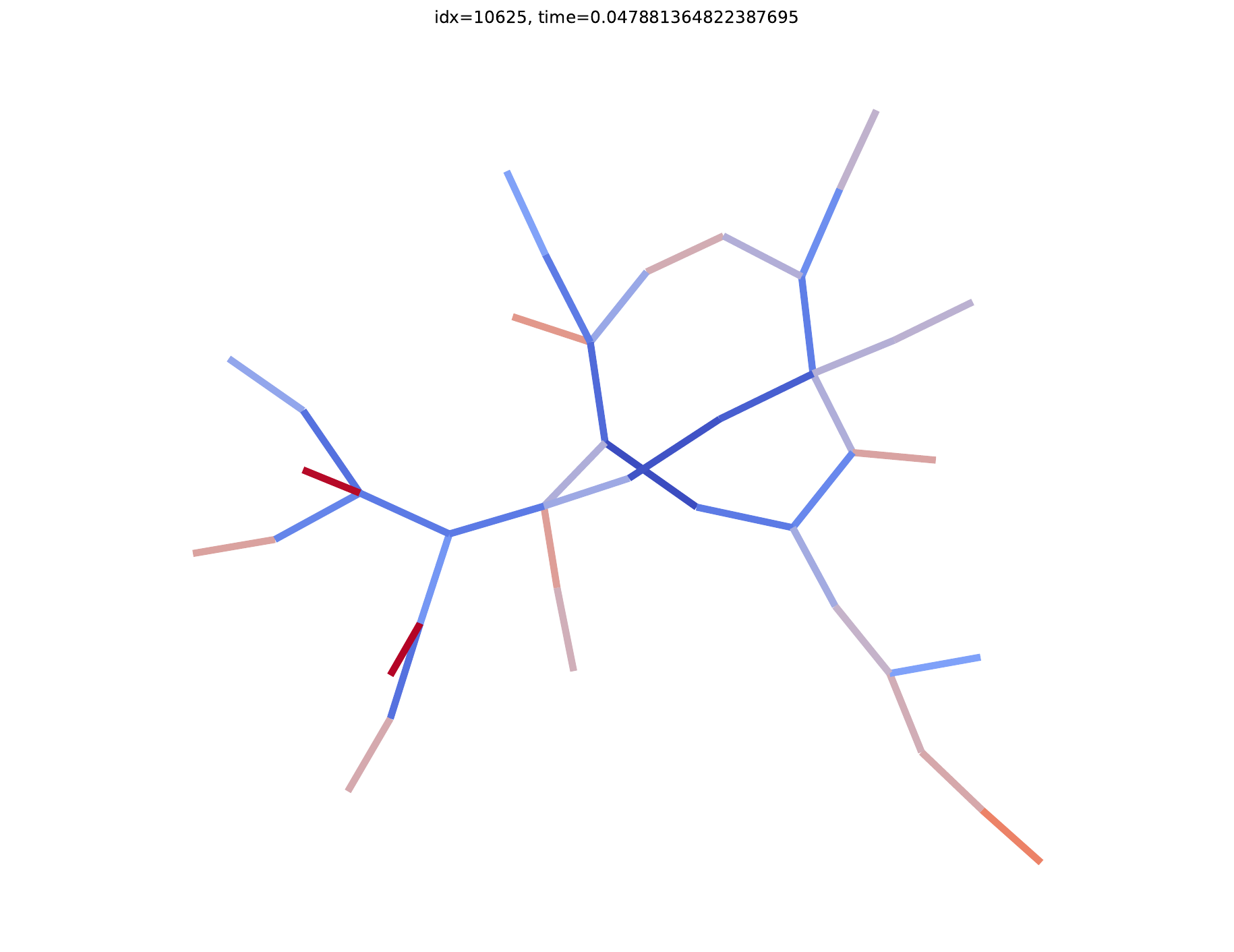} &
\imgcell{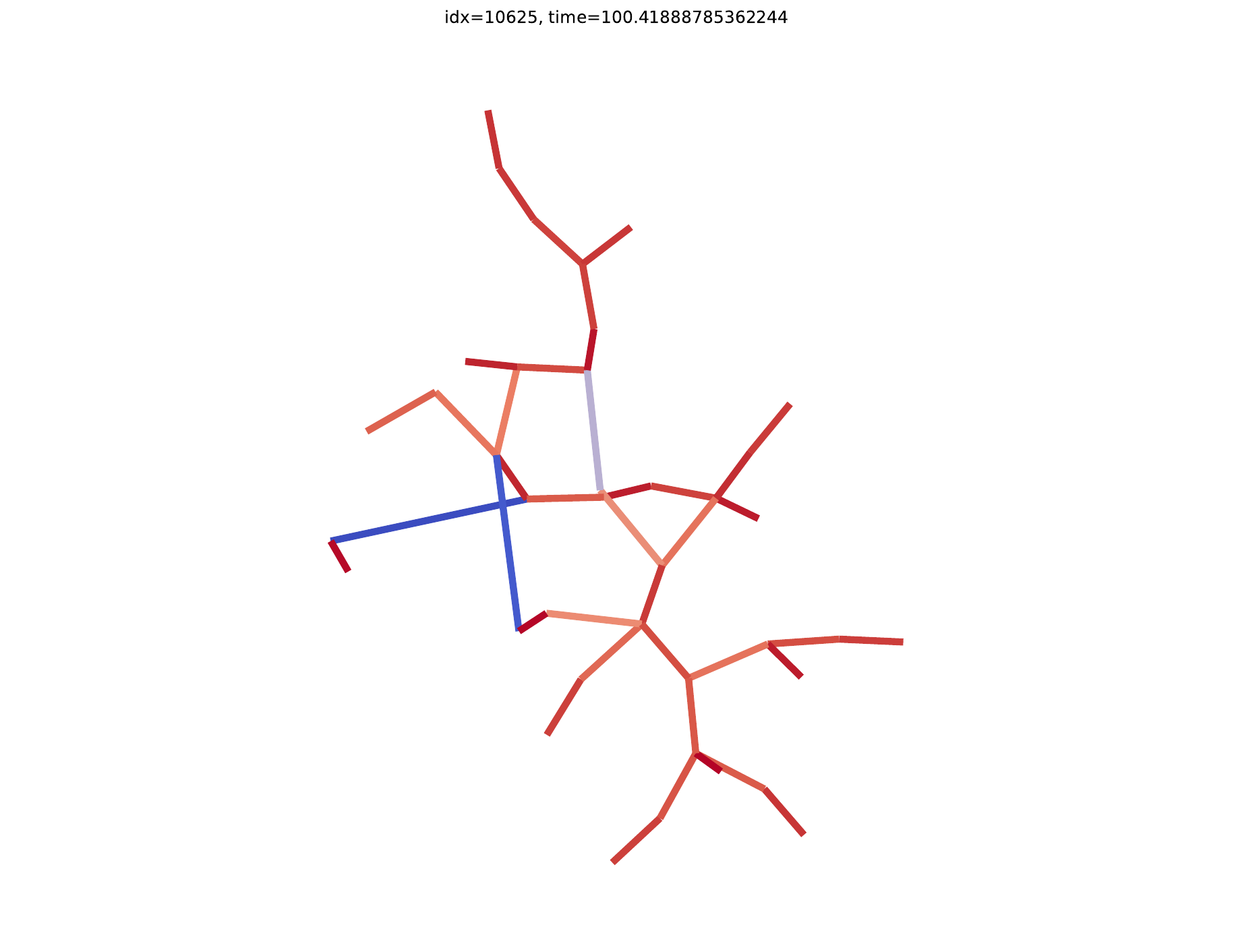} &
\imgcell{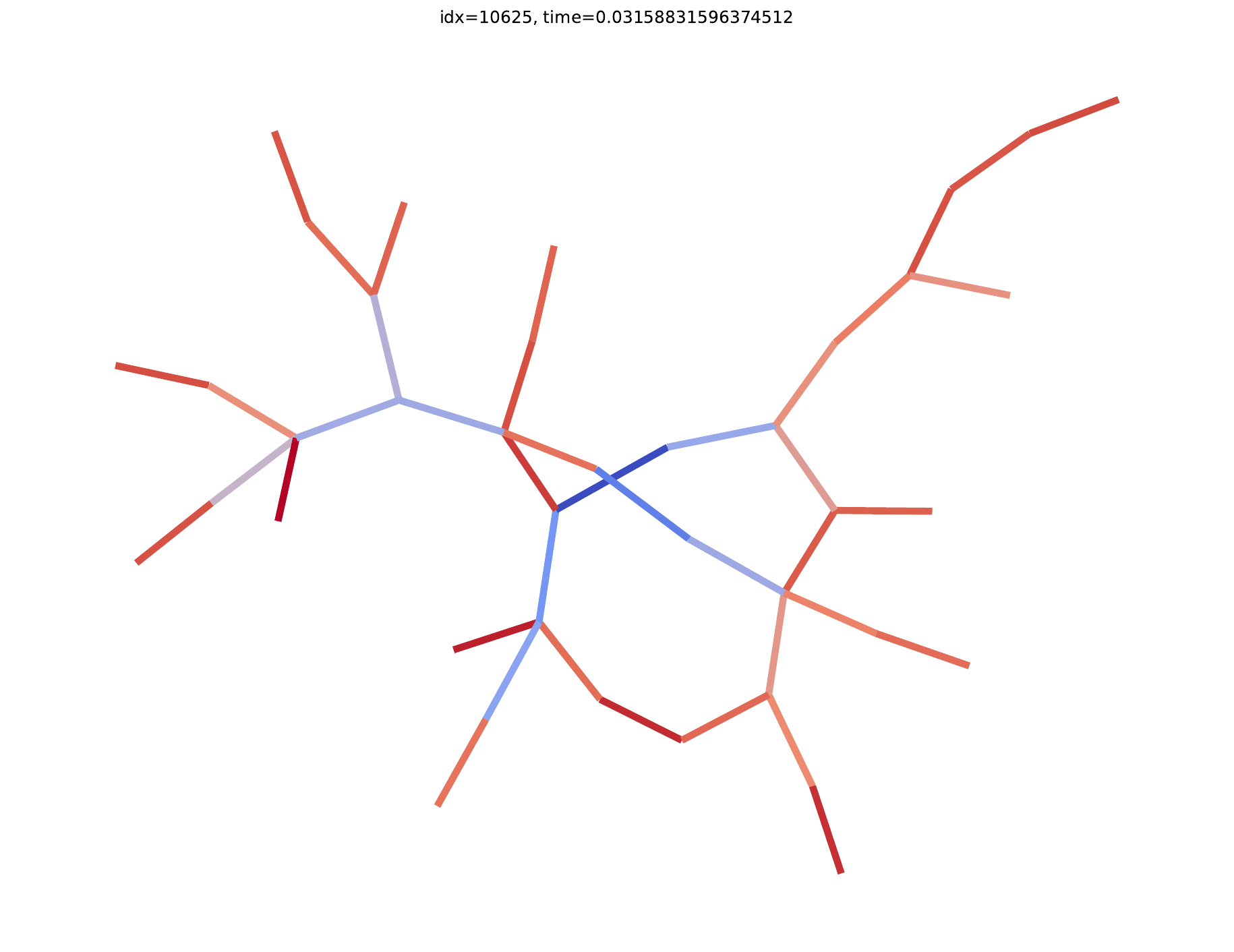} &
\imgcell{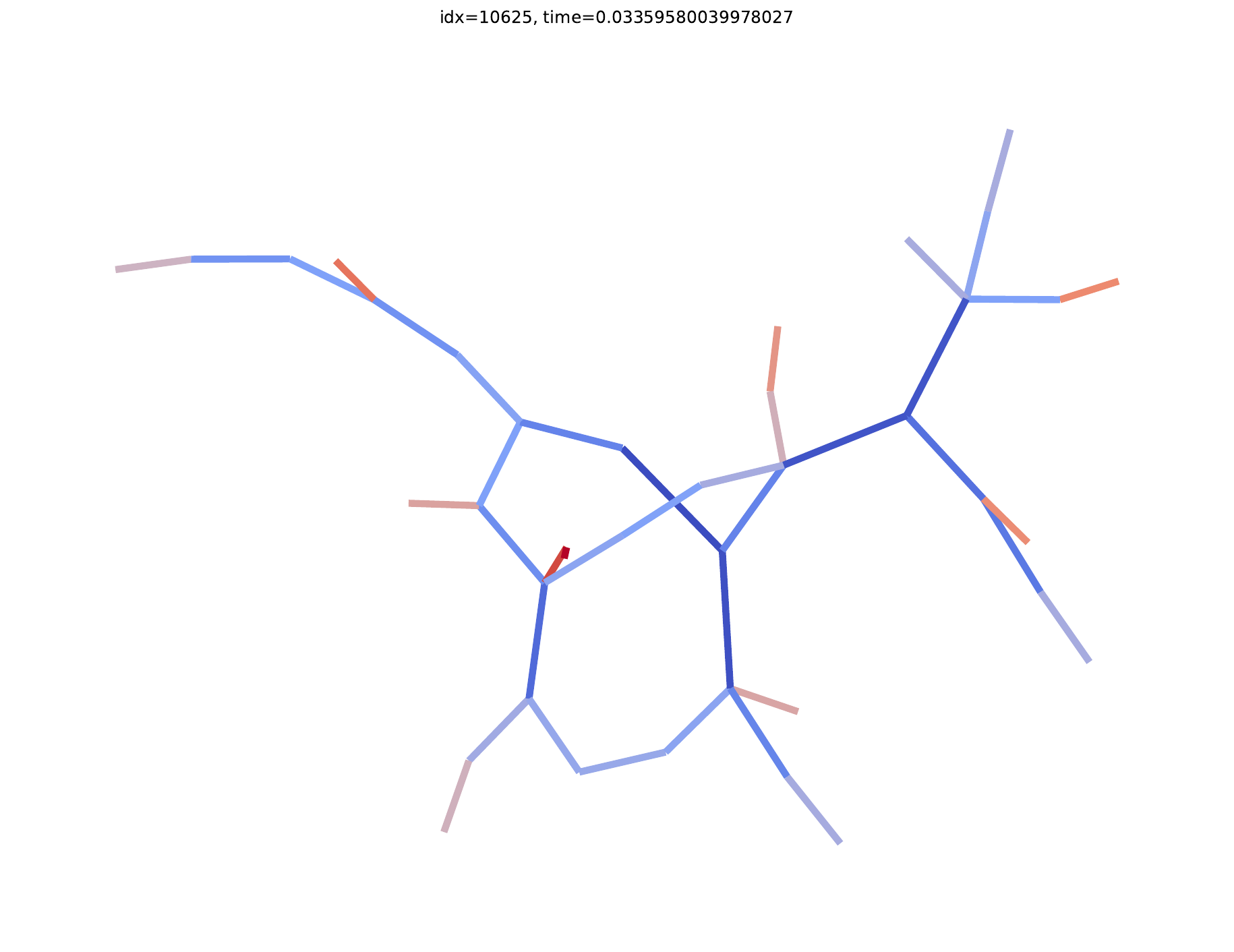} &
\imgcell{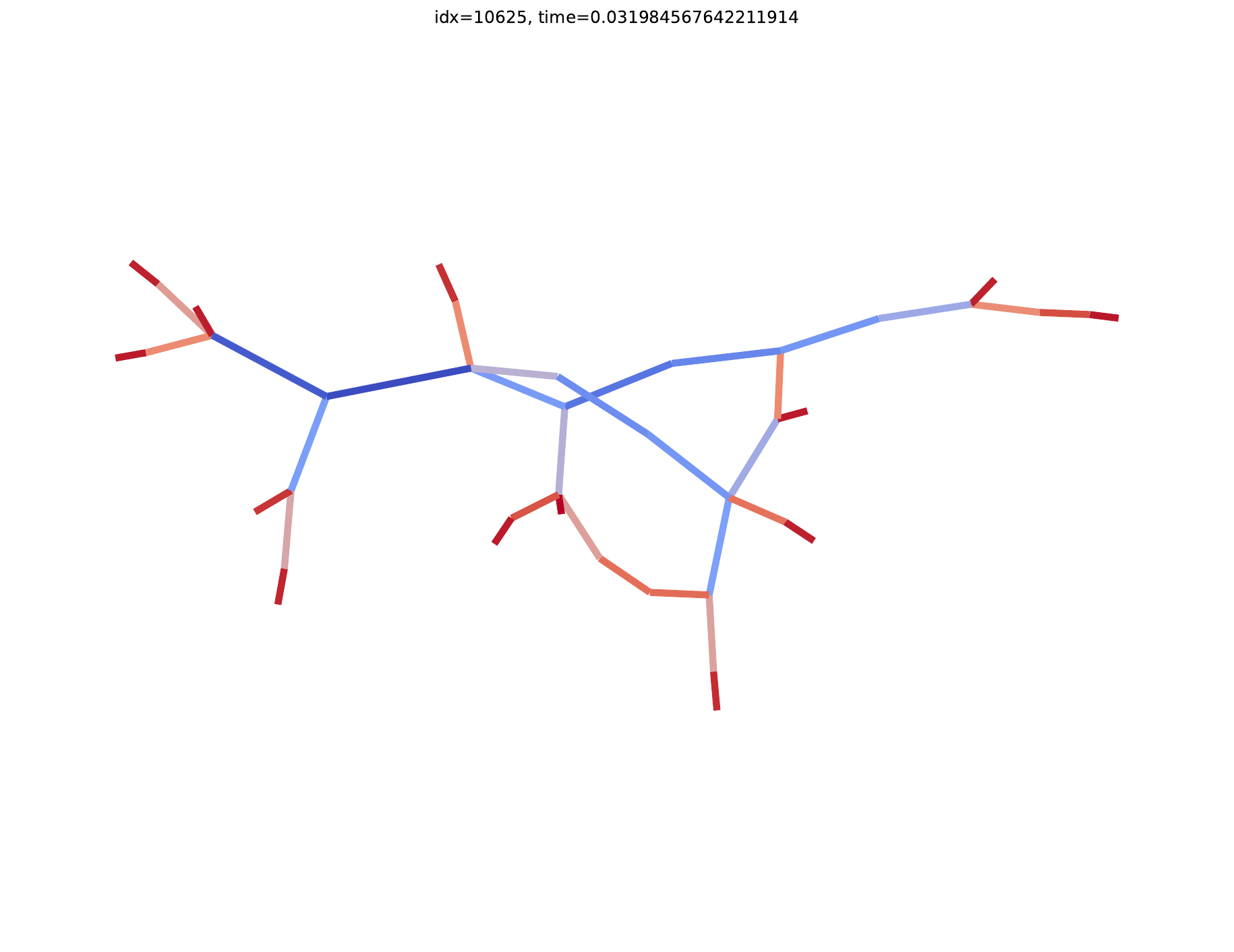} &
\imgcell{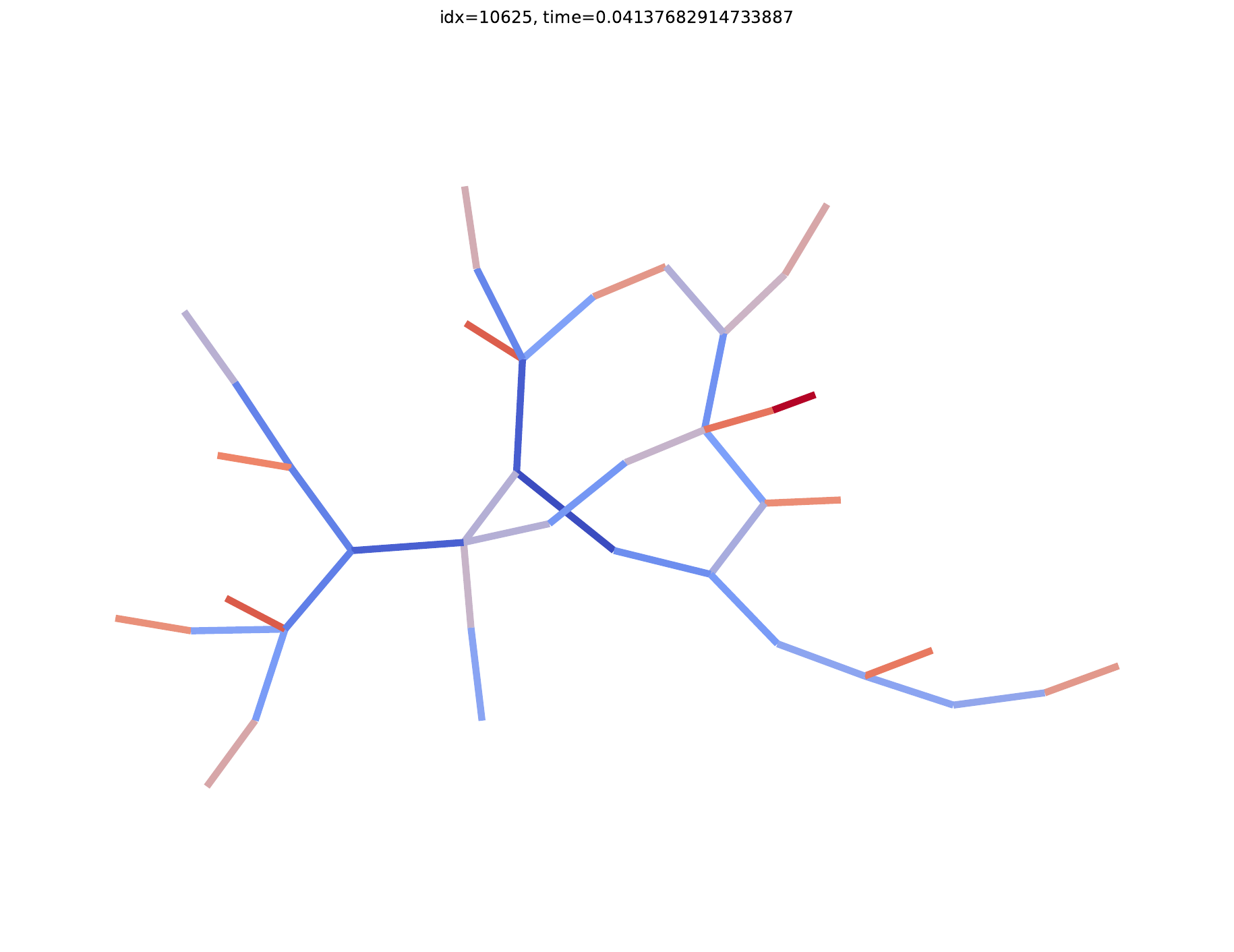} &
\imgcell{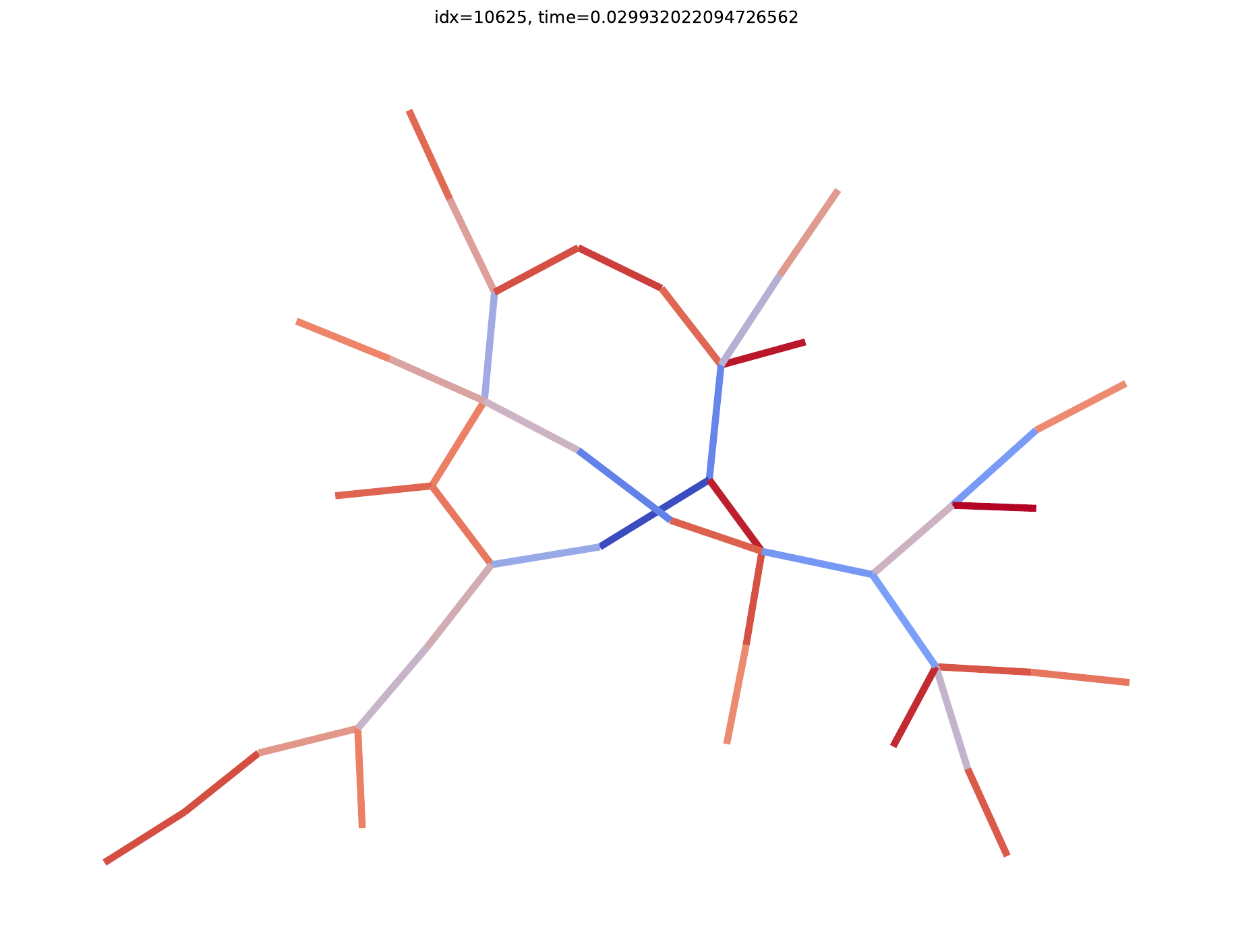} &
\imgcell{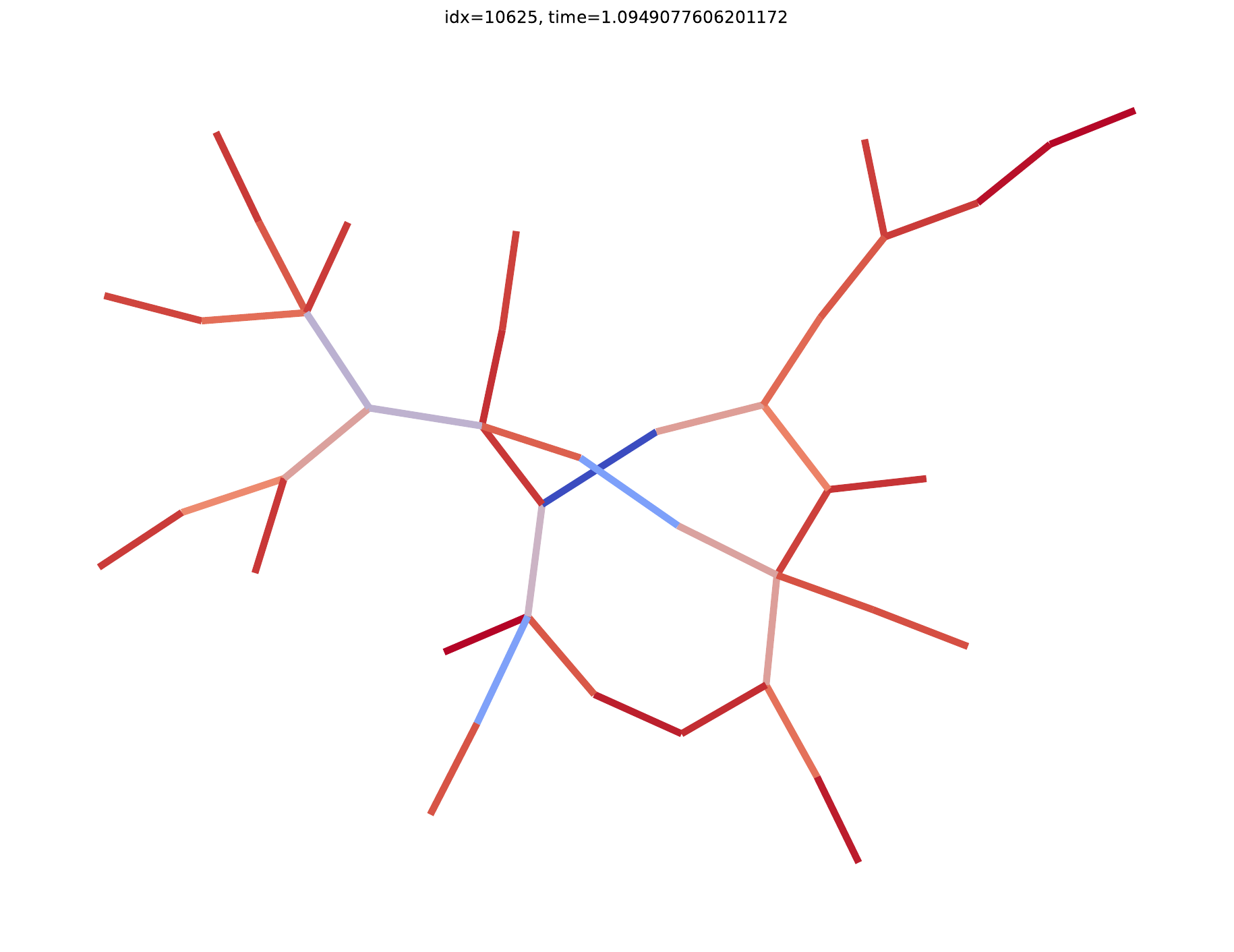} &
\imgcell{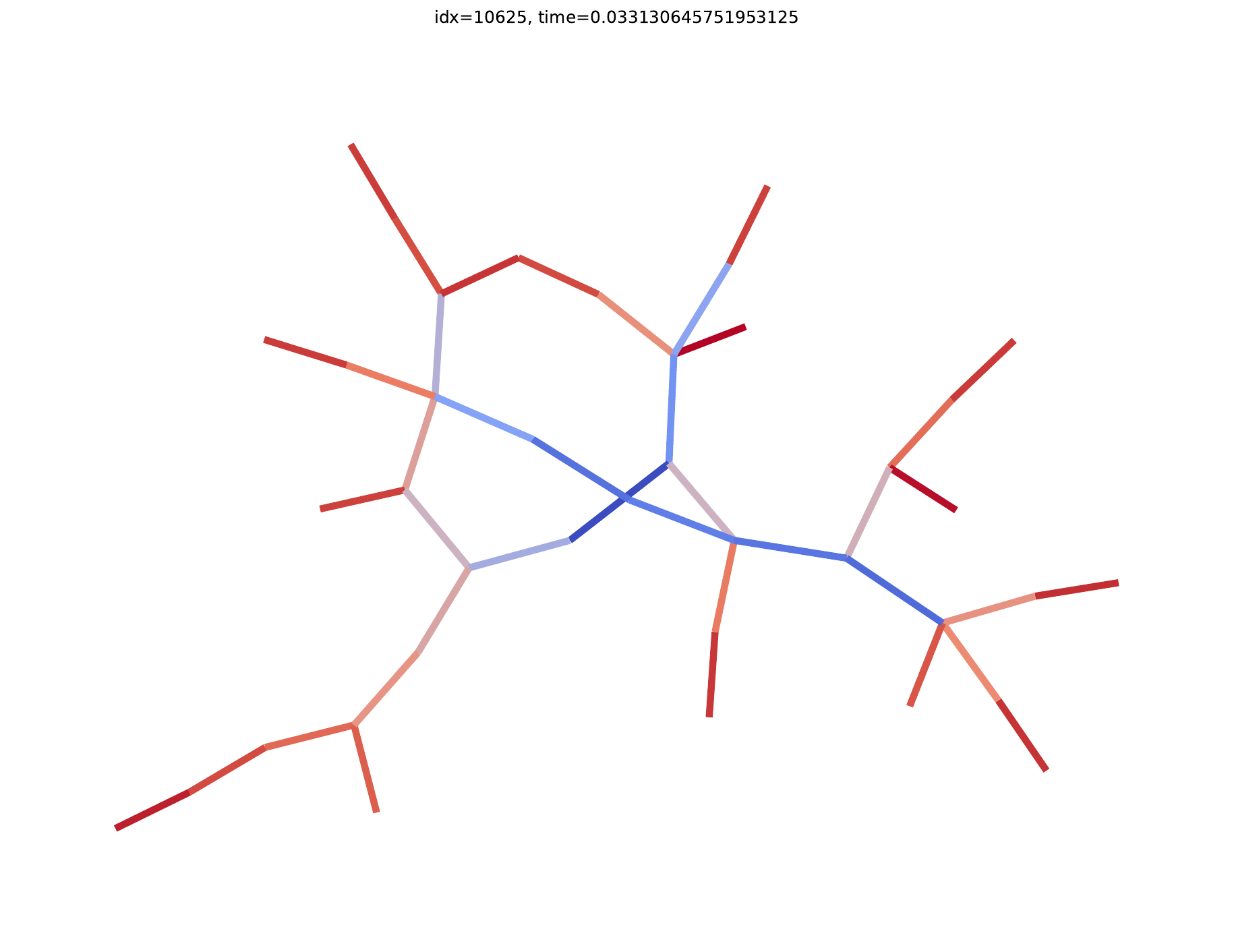} \\

&
t = 0.00s &
t = 0.33s &
t = 0.13s &
t = 0.05s &
t = 100.42s &
t = 0.03s &
t = 0.03s &
t = 0.04s &
t = 0.04s &
t = 0.03s &
t = 0.04s &
t = 0.03s \\

\makecell{\bfseries grafo5759.35\\N = 40\\M = 50} &
\imgcell{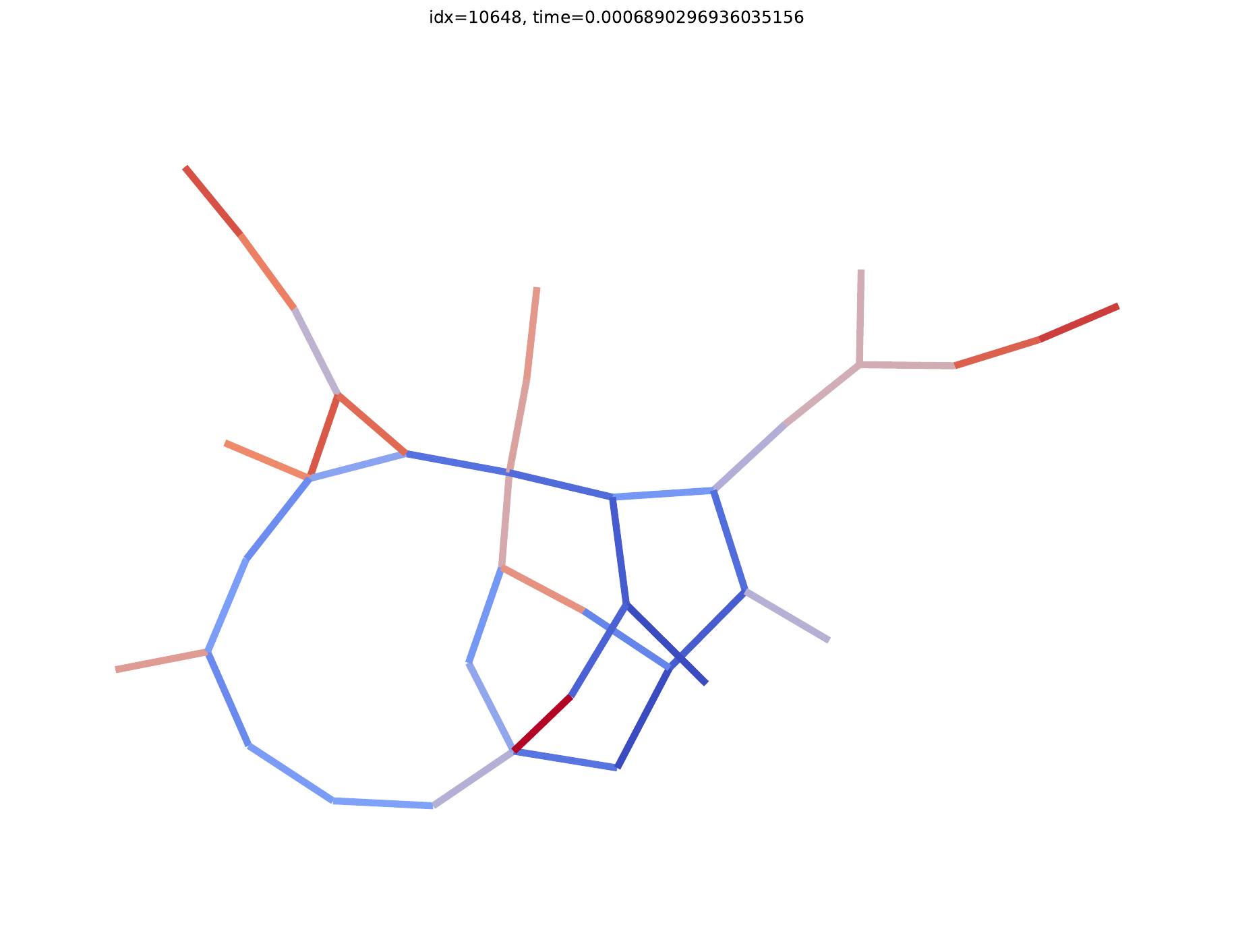} &
\imgcell{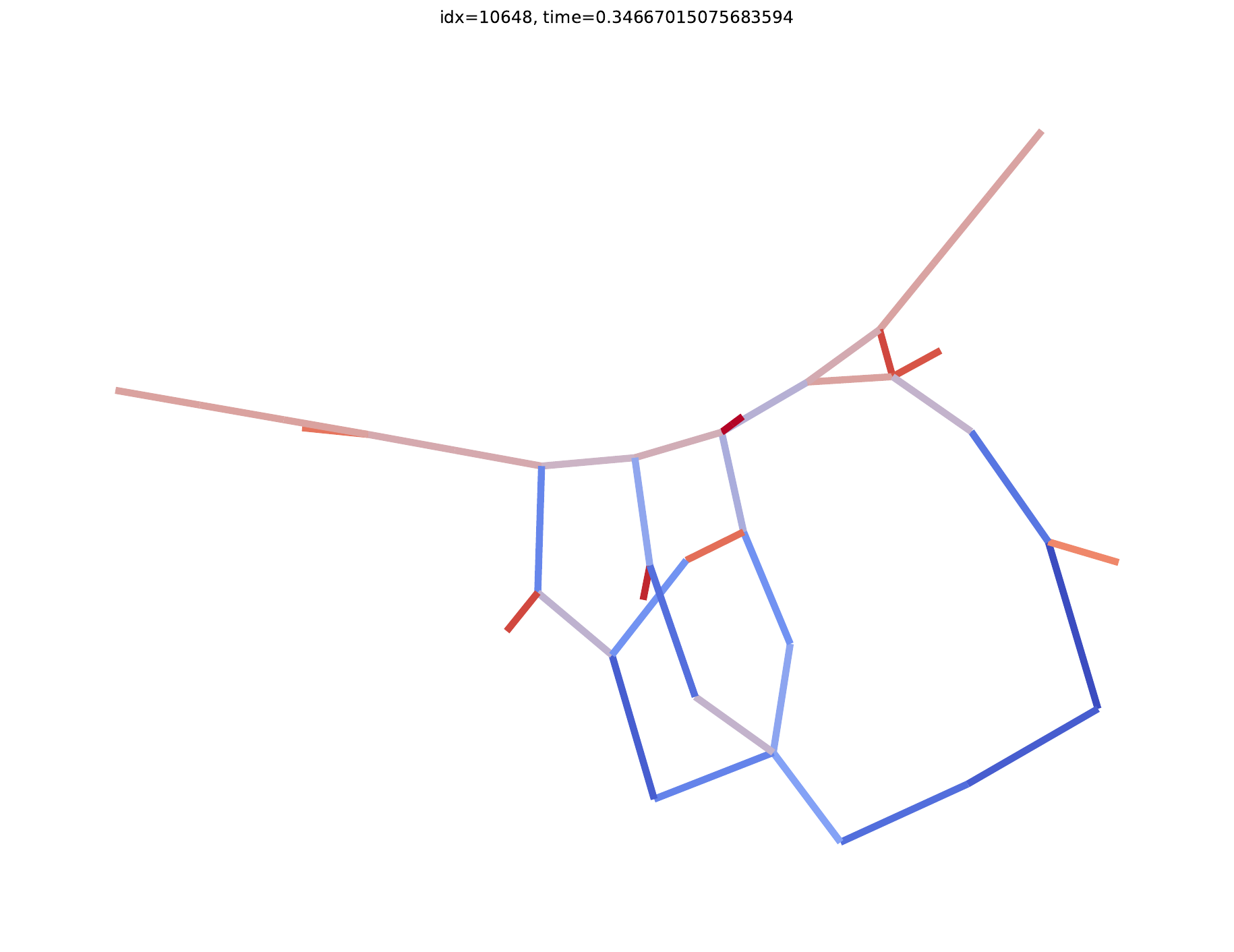} &
\imgcell{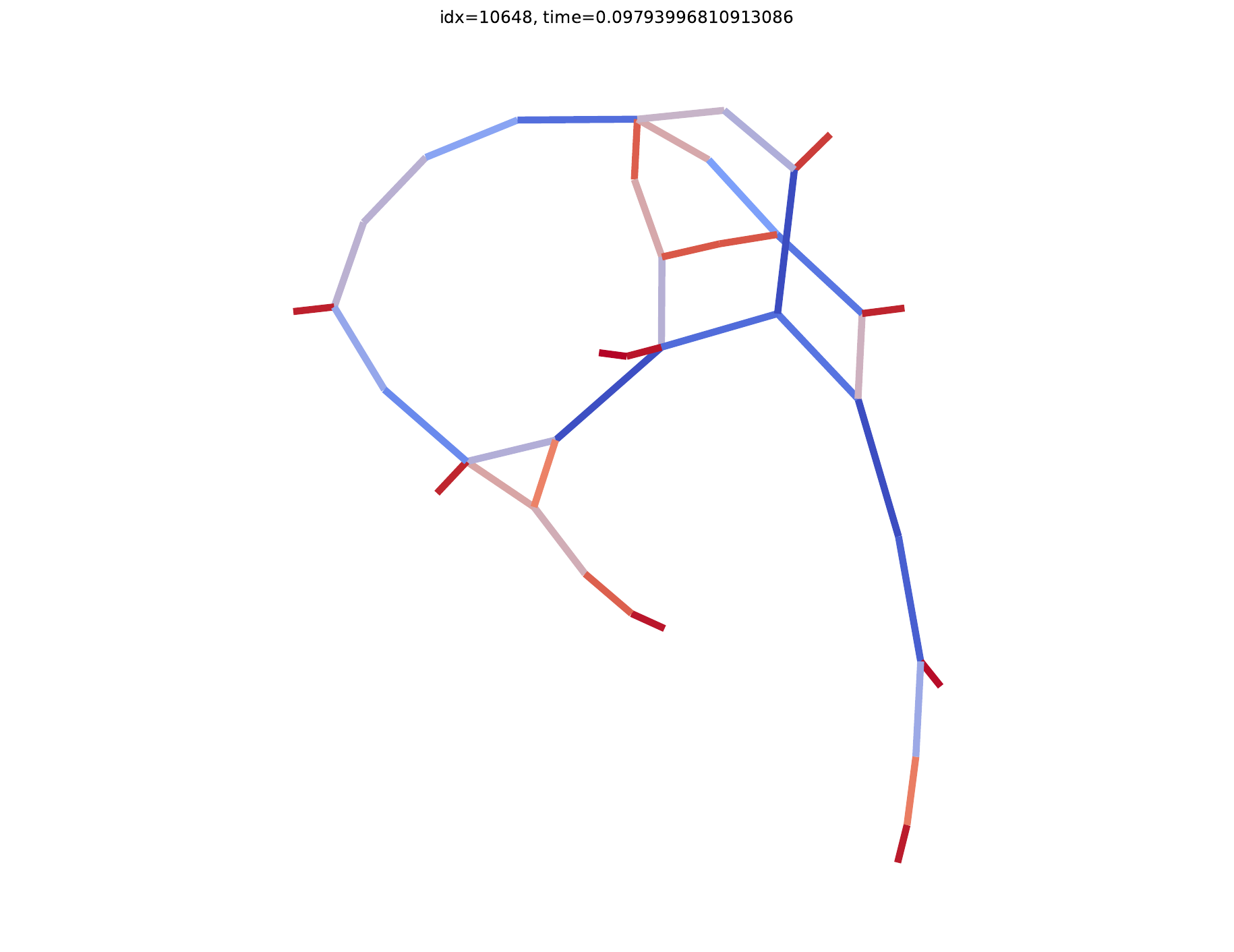} &
\imgcell{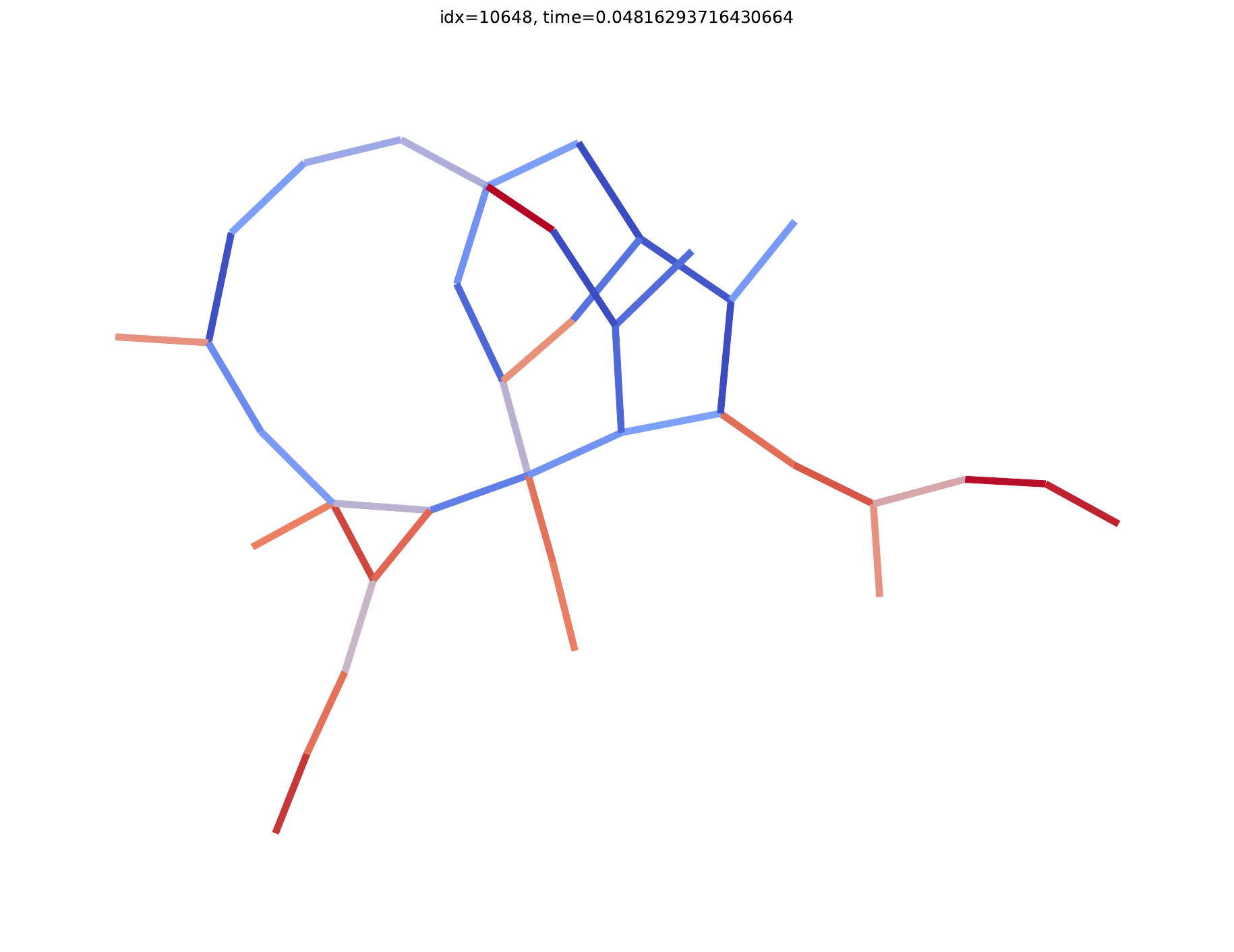} &
\imgcell{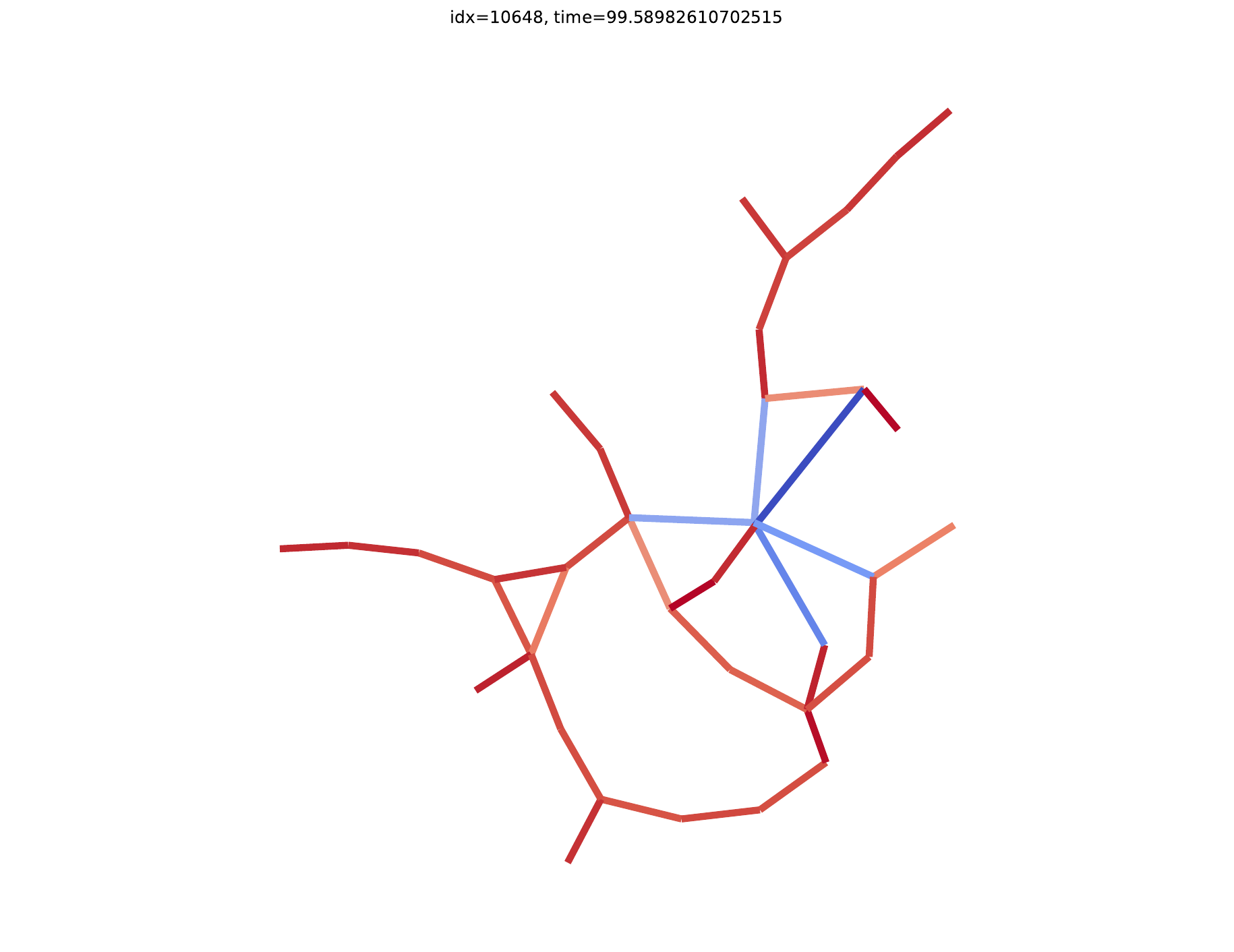} &
\imgcell{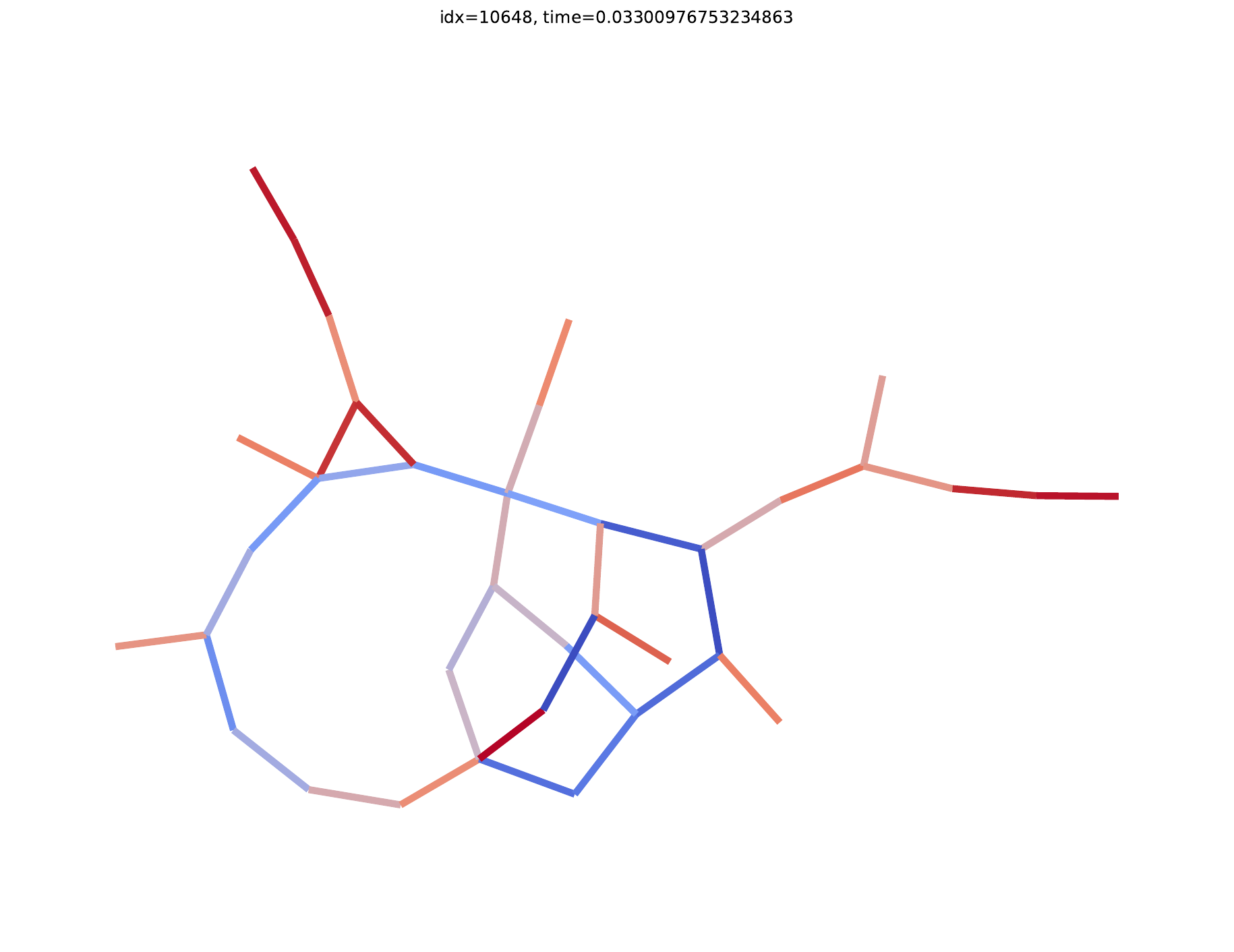} &
\imgcell{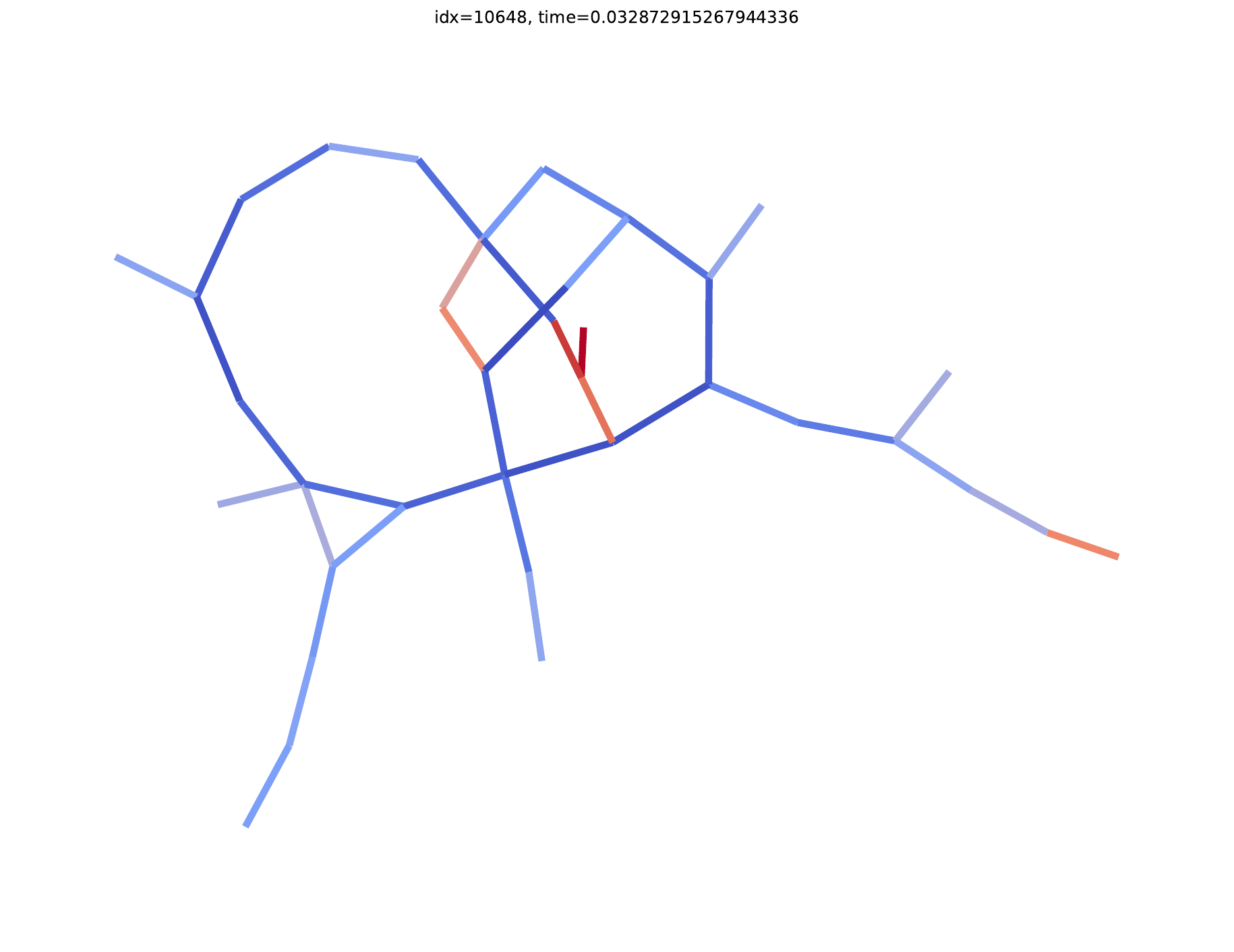} &
\imgcell{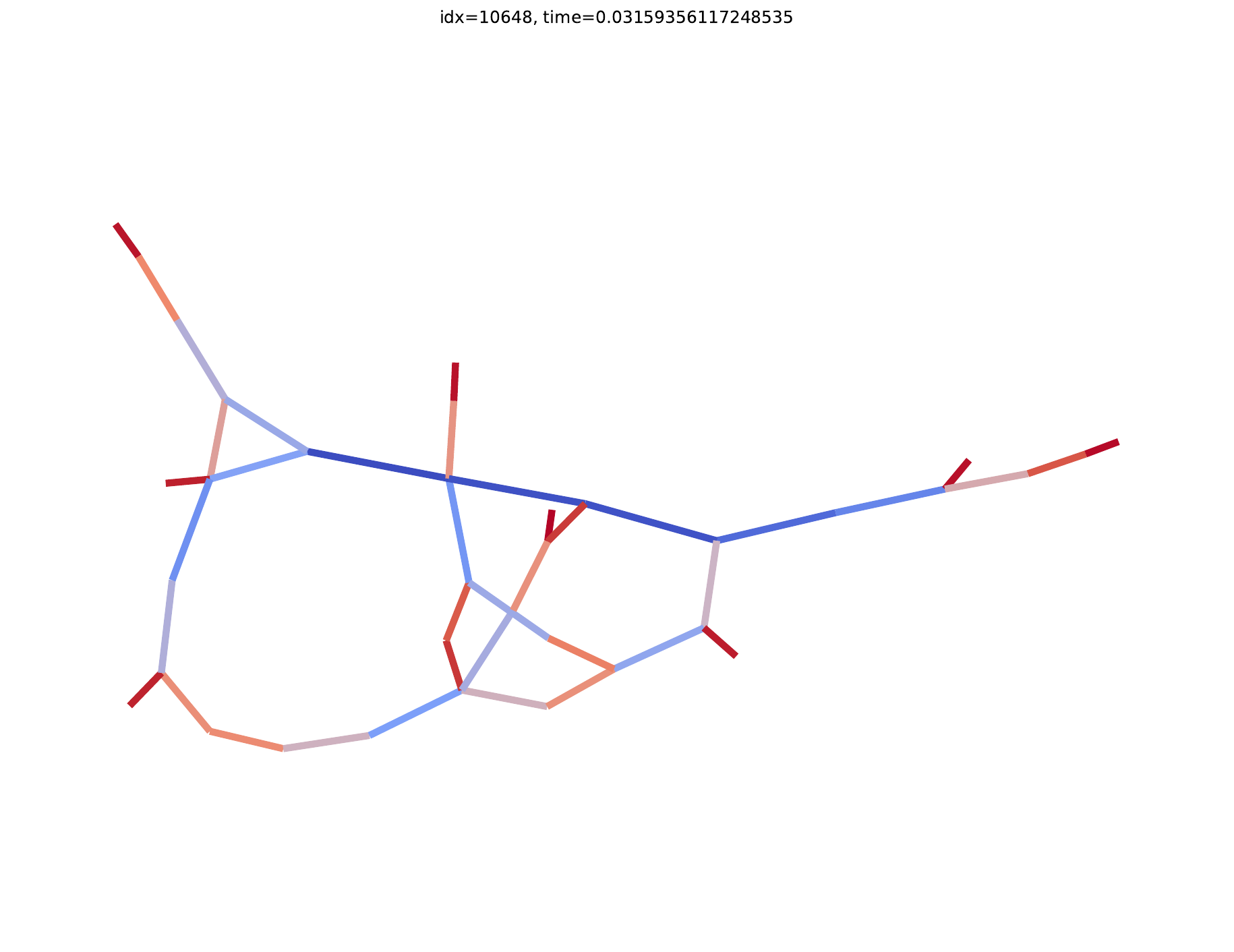} &
\imgcell{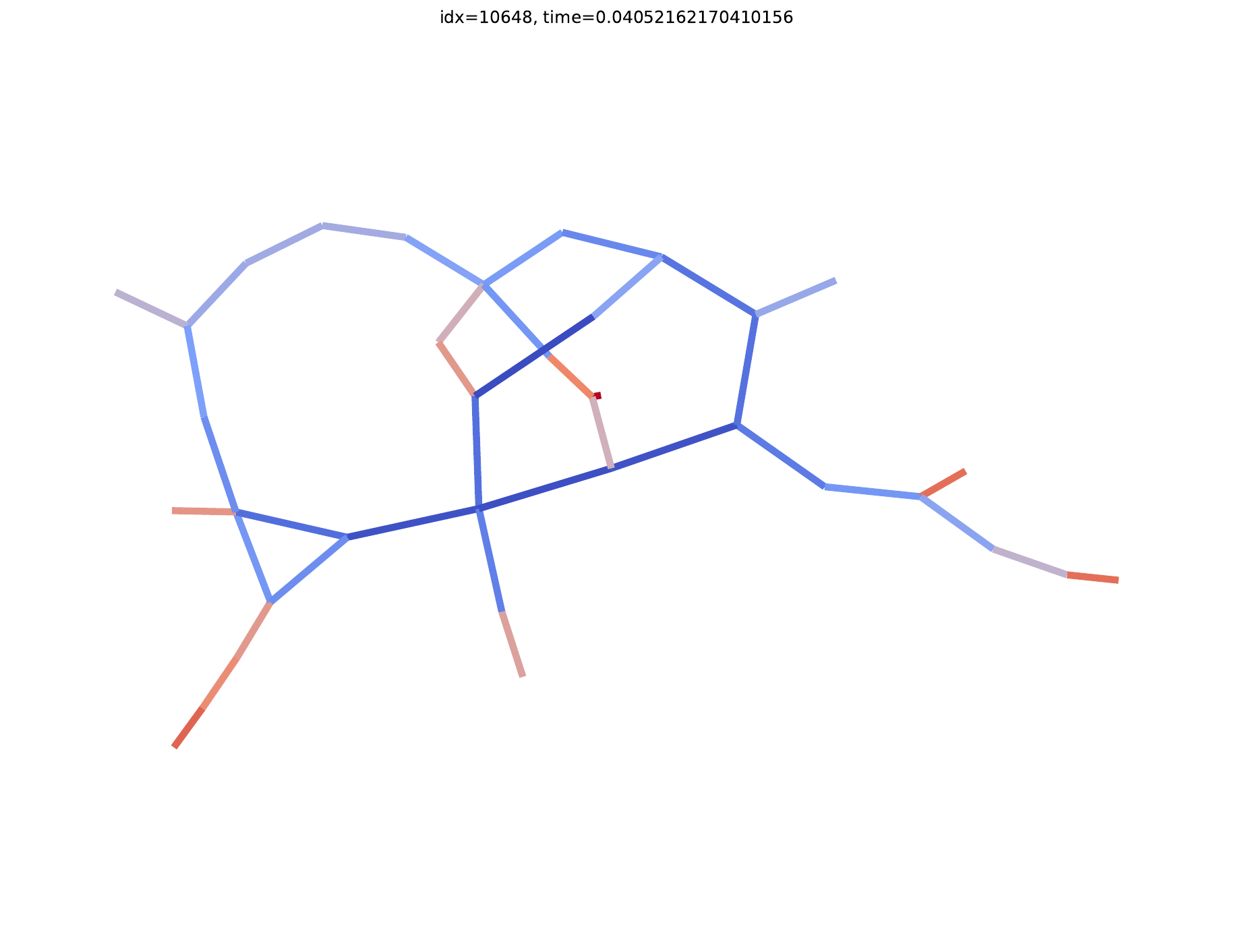} &
\imgcell{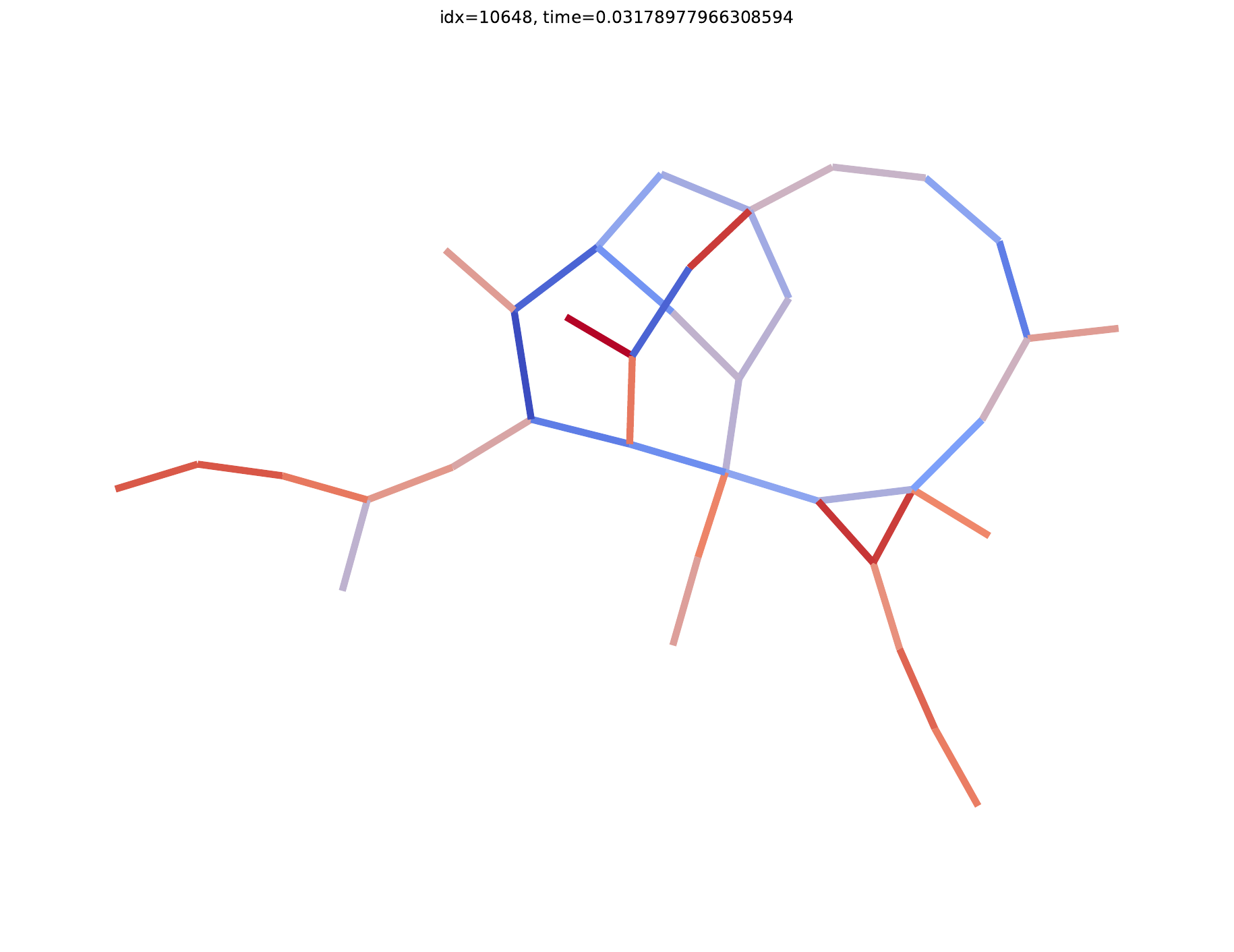} &
\imgcell{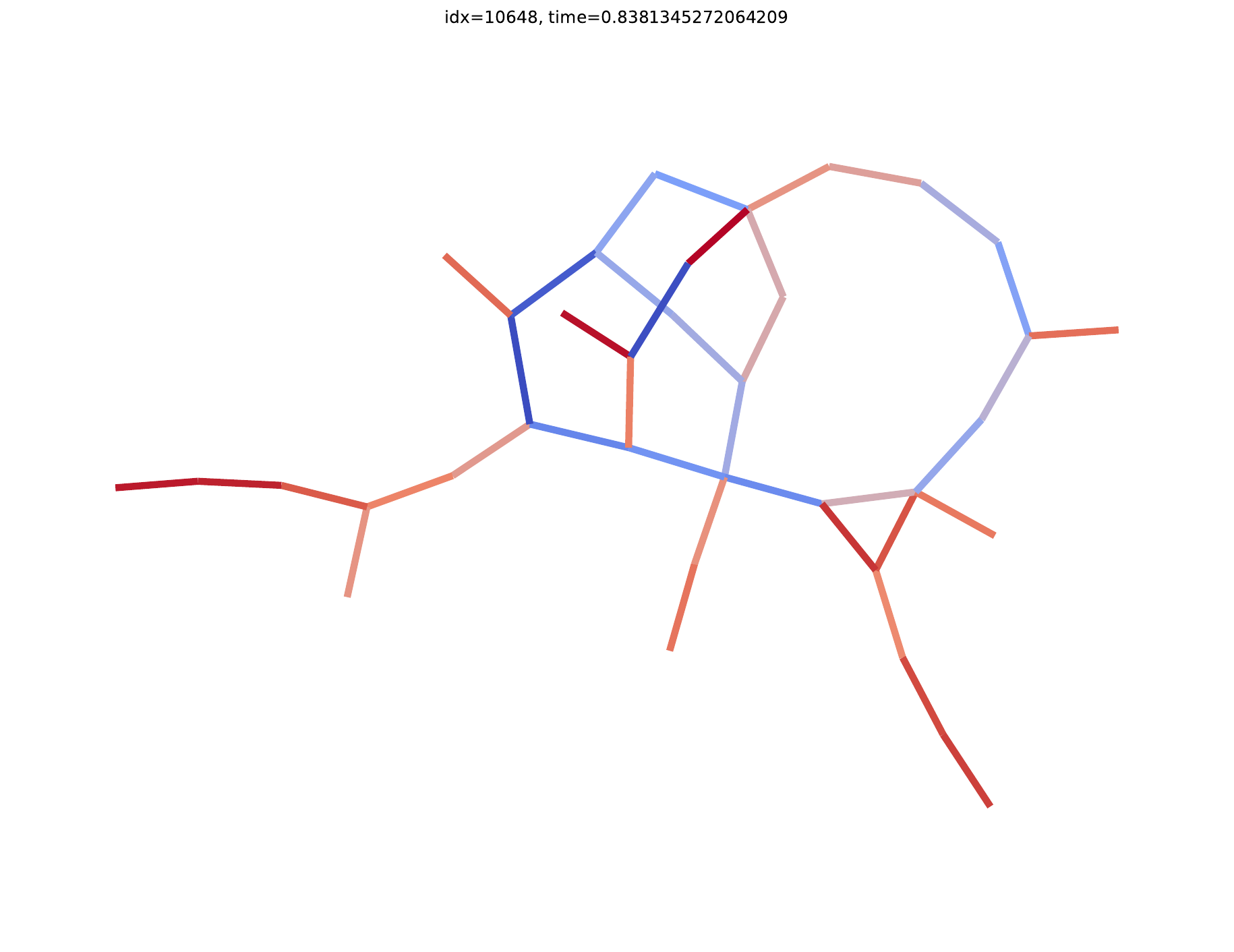} &
\imgcell{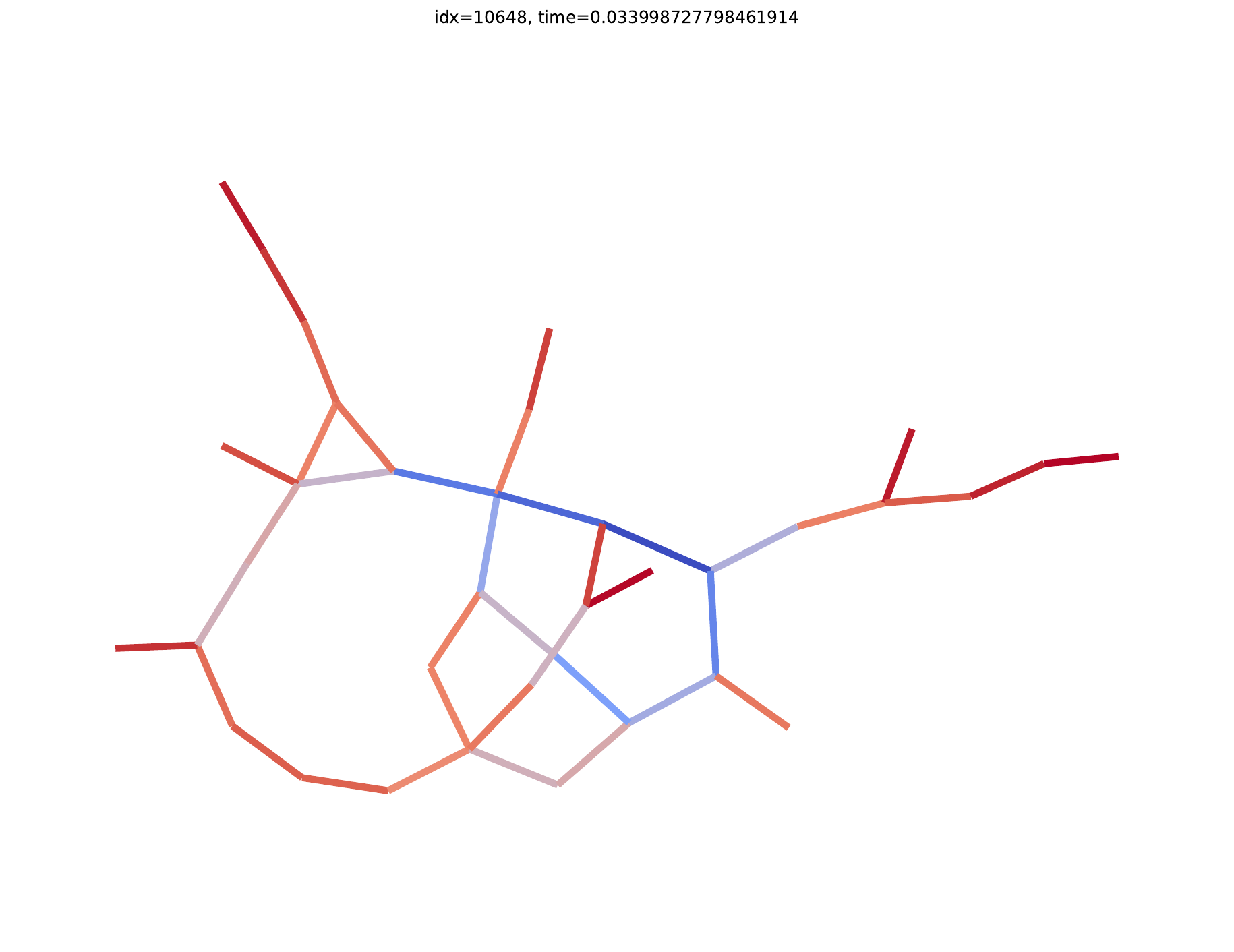} \\

&
t = 0.00s &
t = 0.35s &
t = 0.10s &
t = 0.05s &
t = 99.59s &
t = 0.03s &
t = 0.03s &
t = 0.05s &
t = 0.04s &
t = 0.03s &
t = 0.04s &
t = 0.03s \\

\makecell{\bfseries grafo2066.33\\N = 37\\M = 43} &
\imgcell{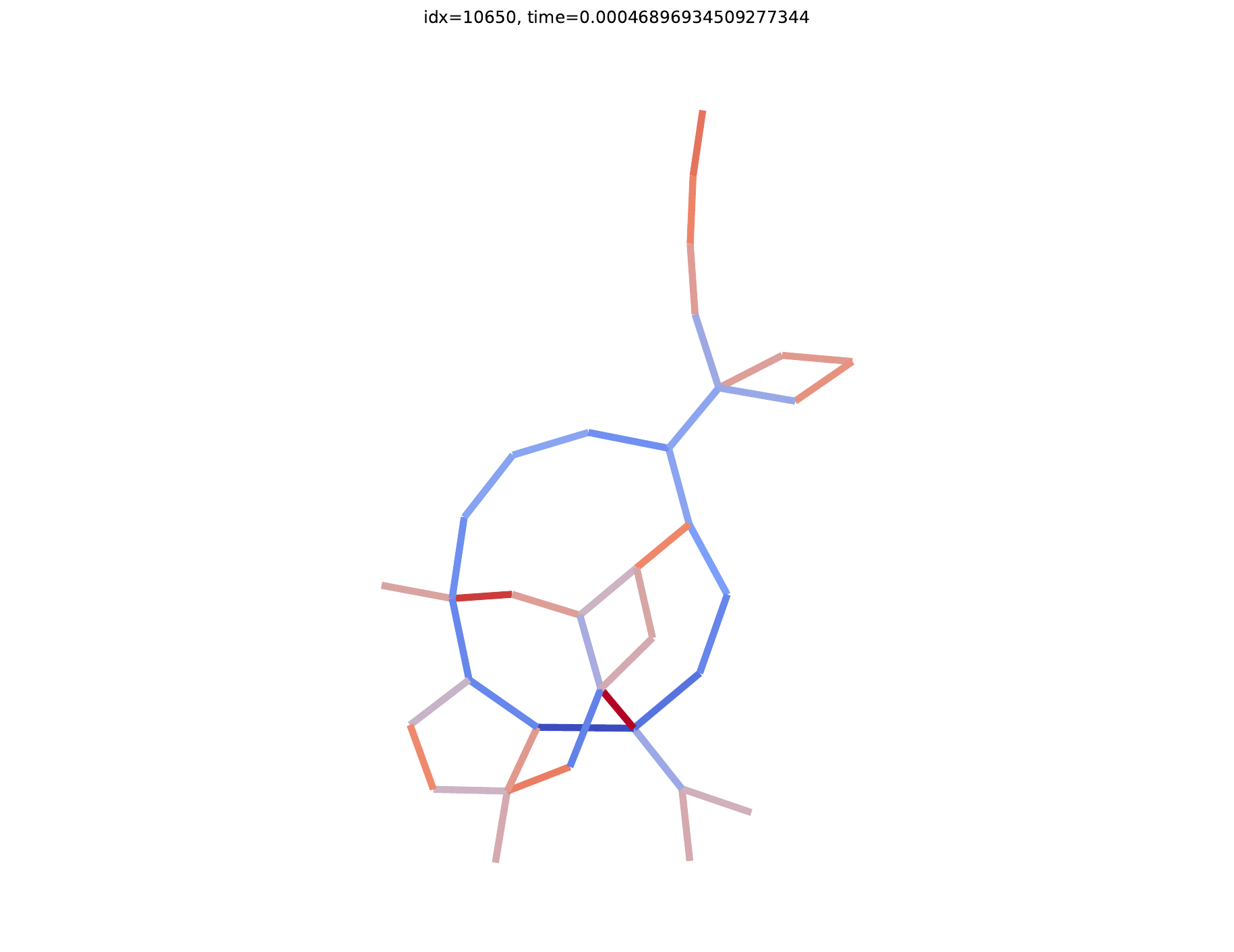} &
\imgcell{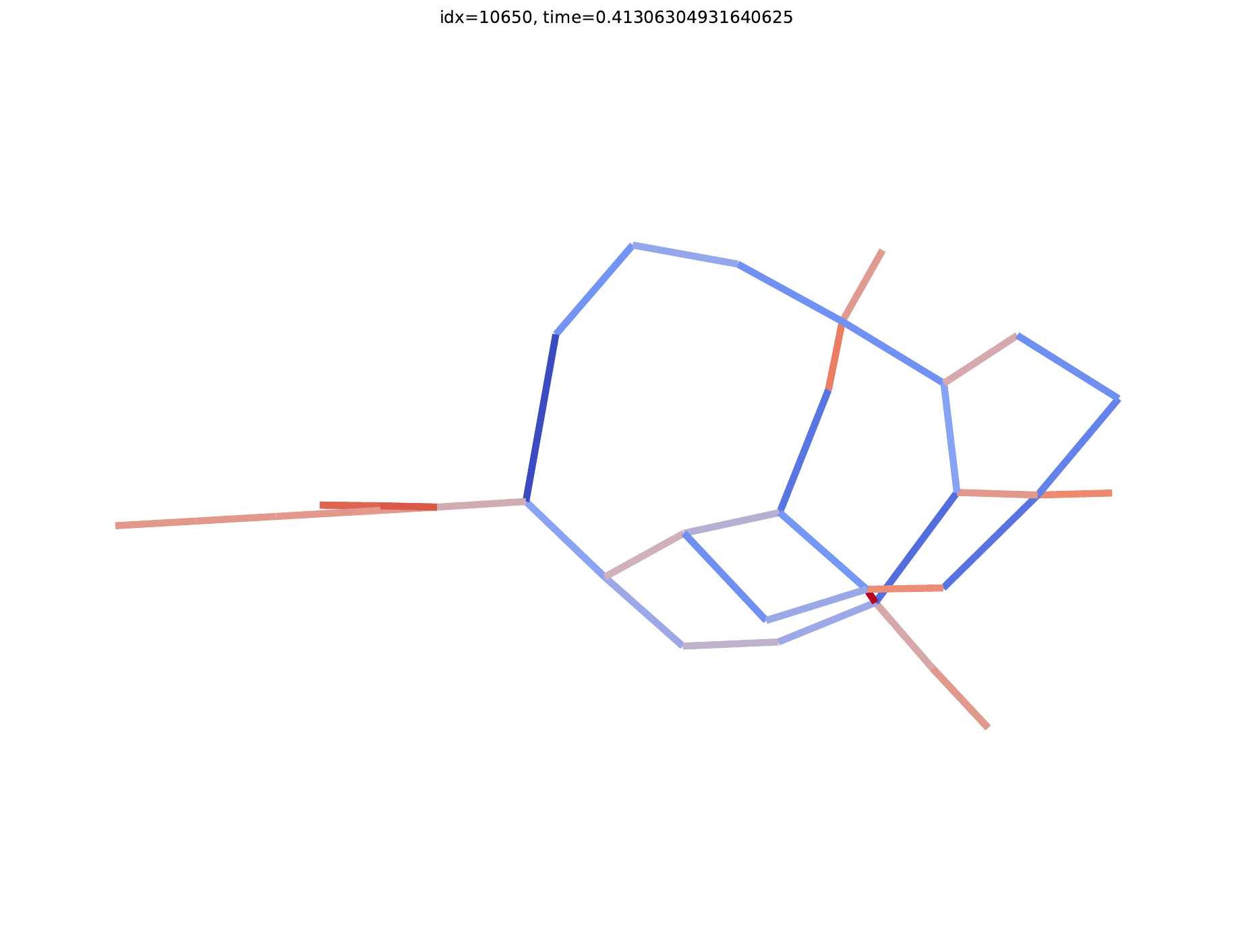} &
\imgcell{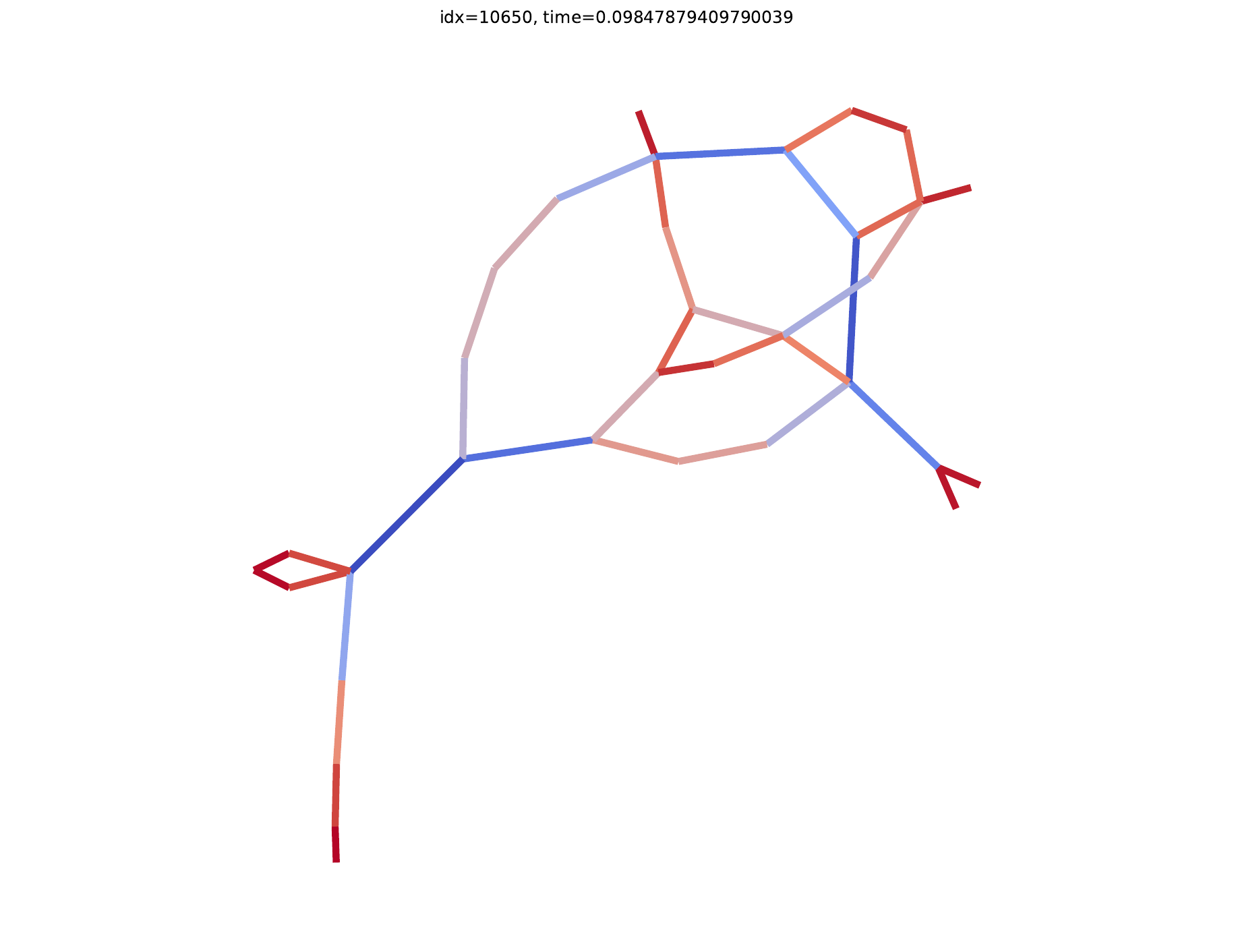} &
\imgcell{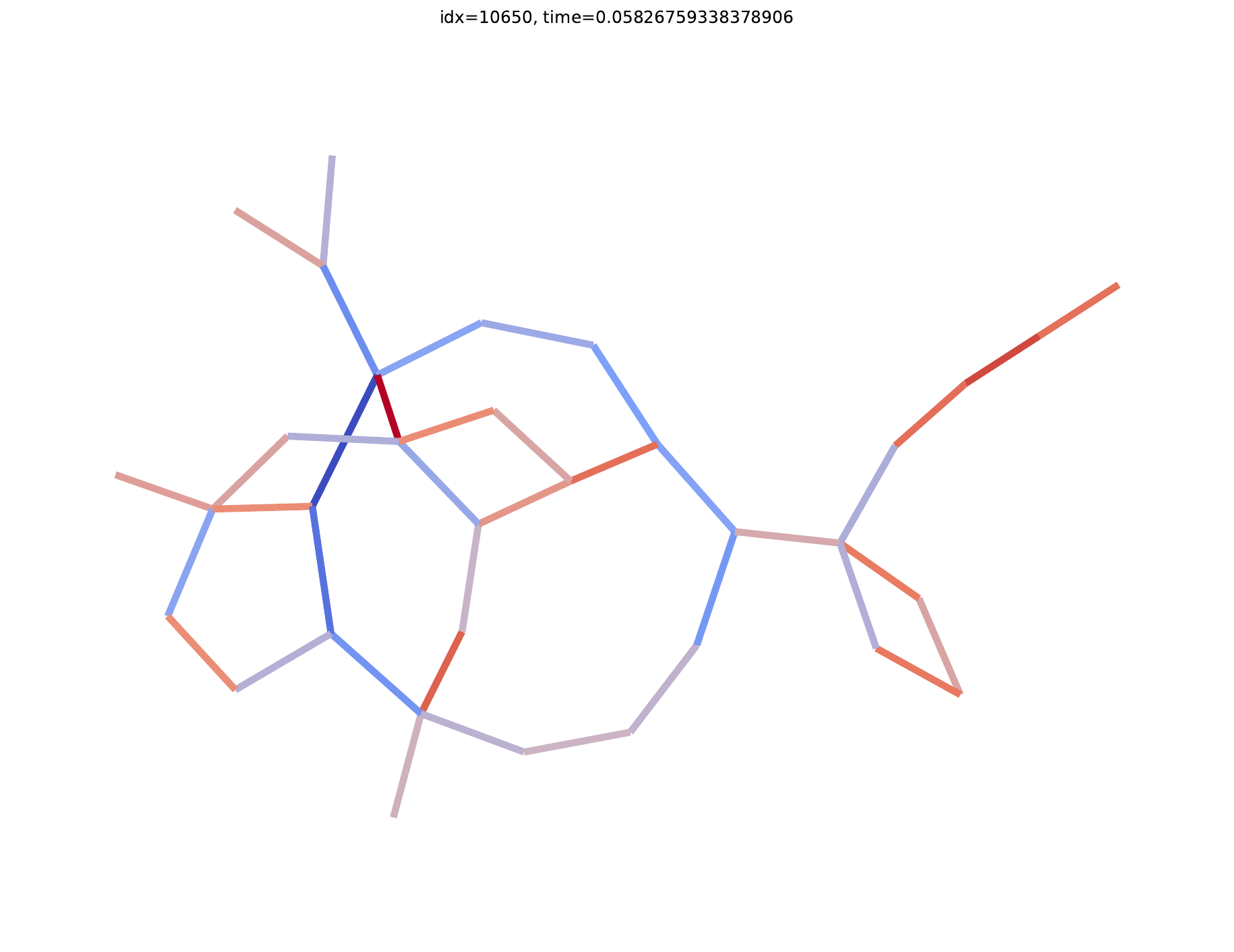} &
\imgcell{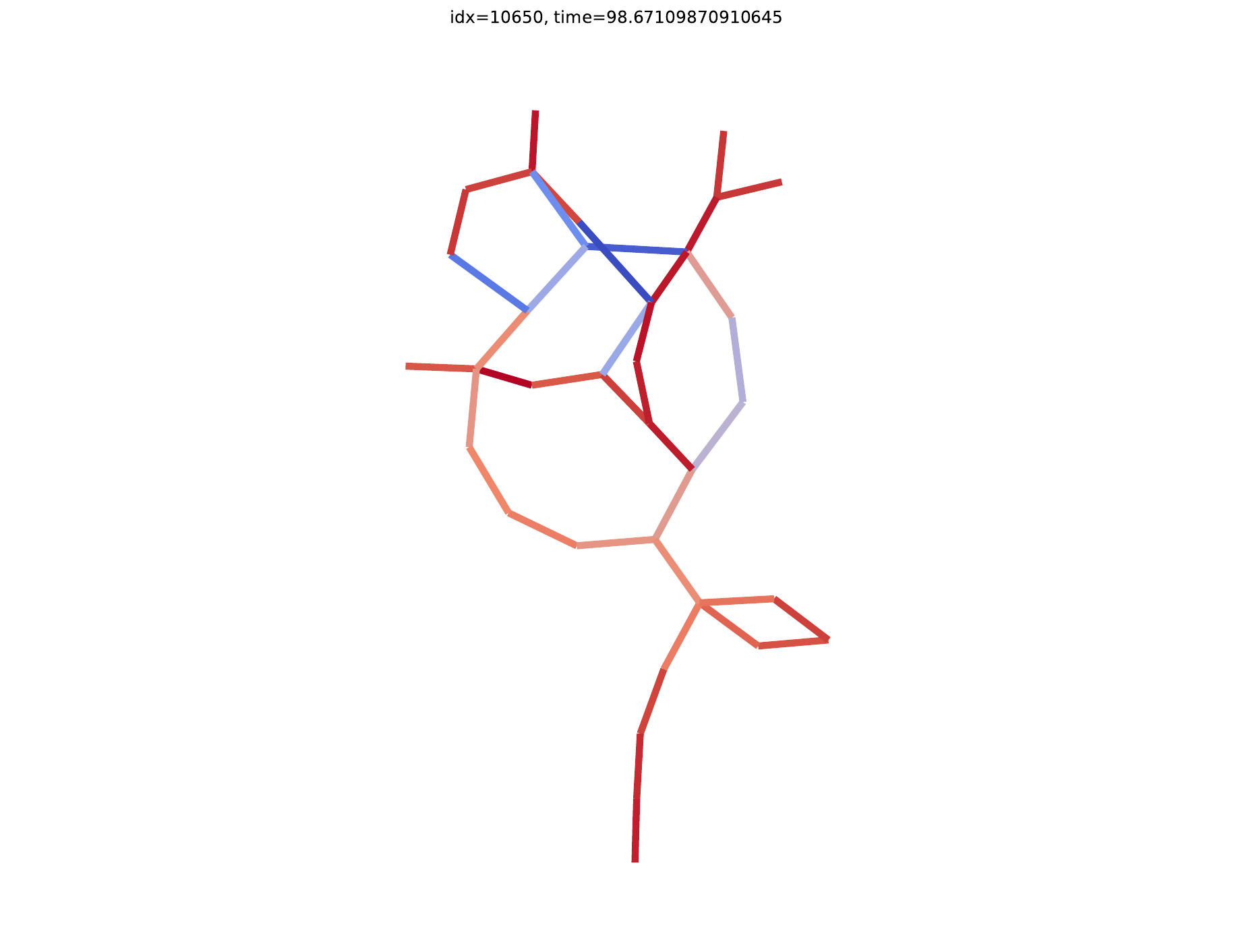} &
\imgcell{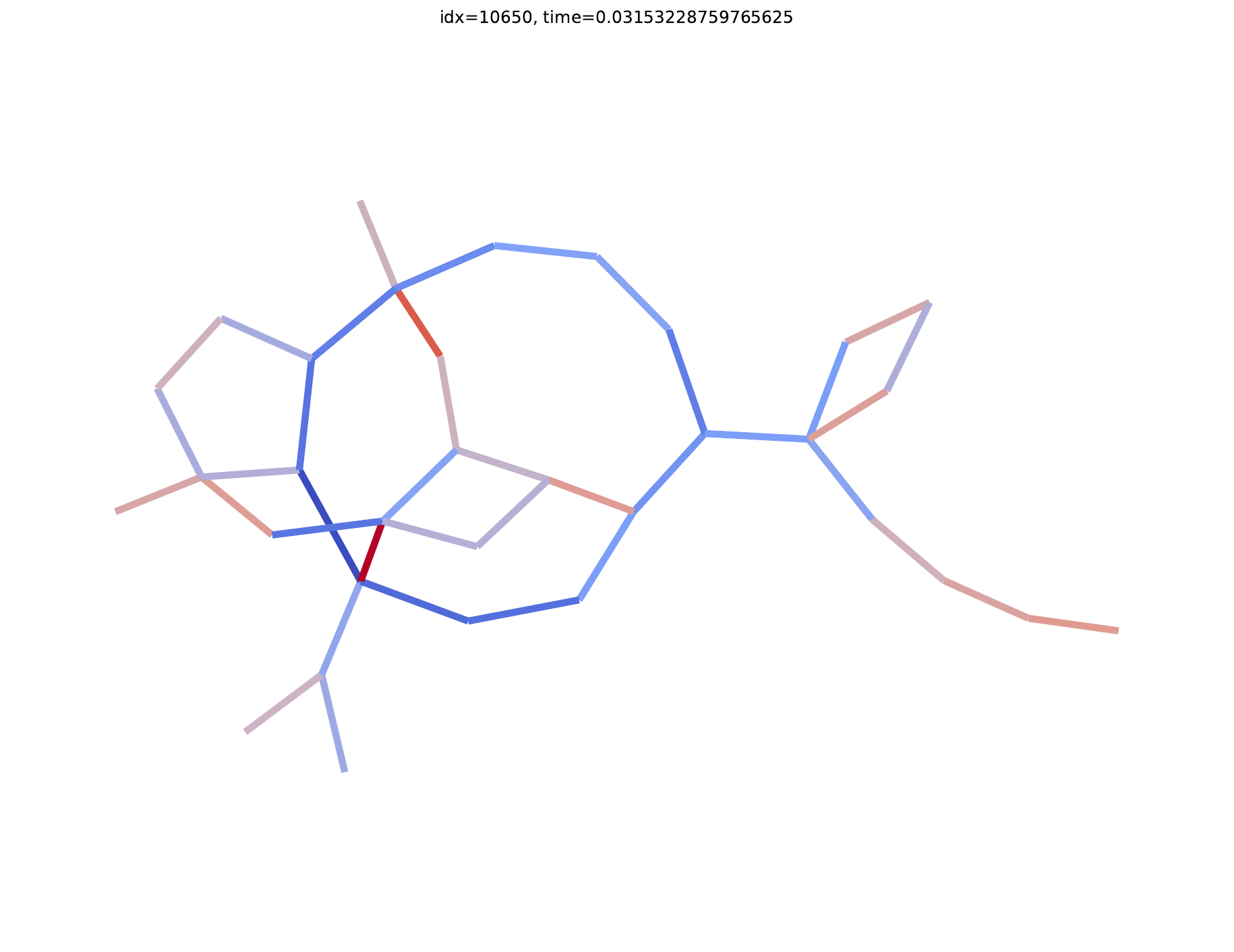} &
\imgcell{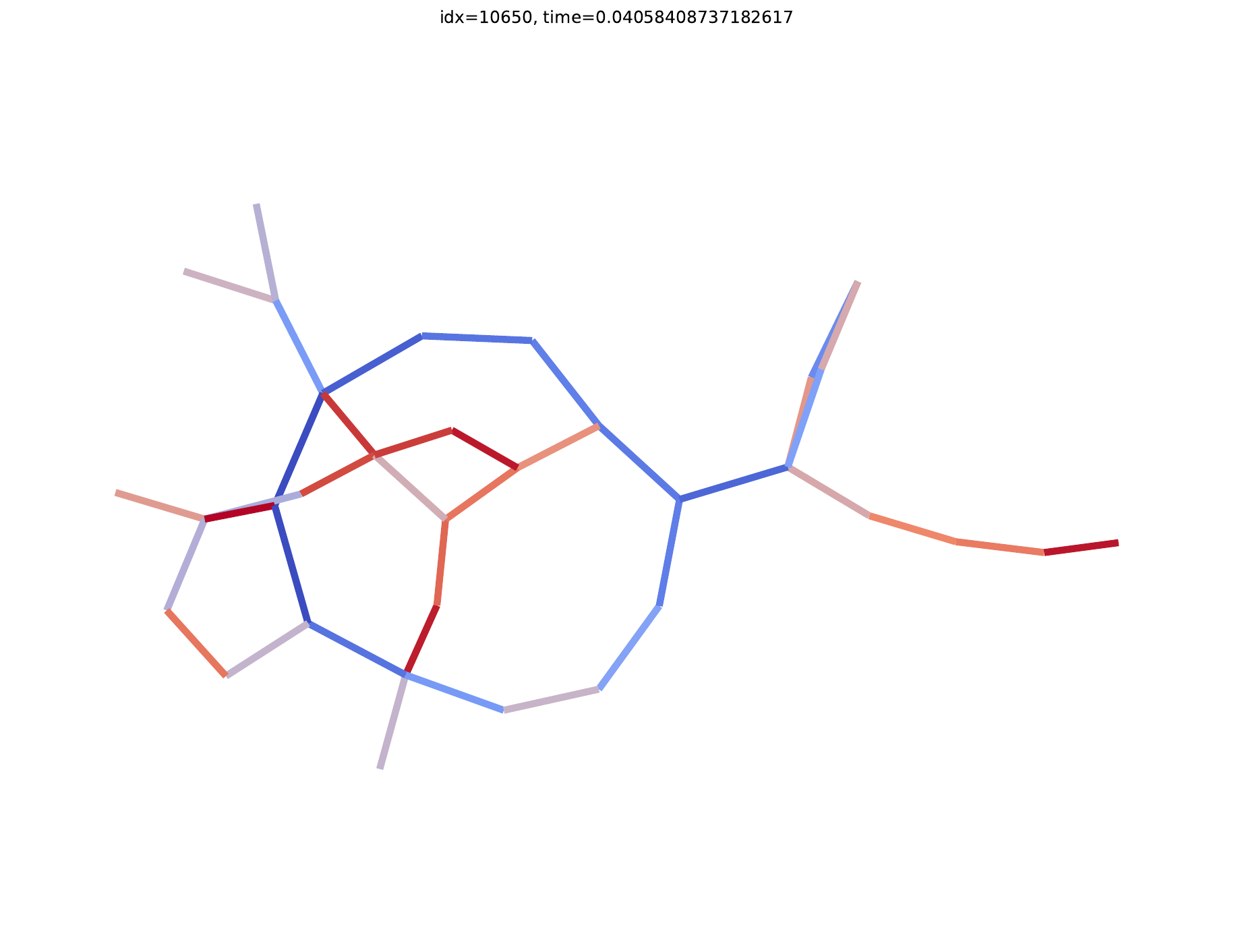} &
\imgcell{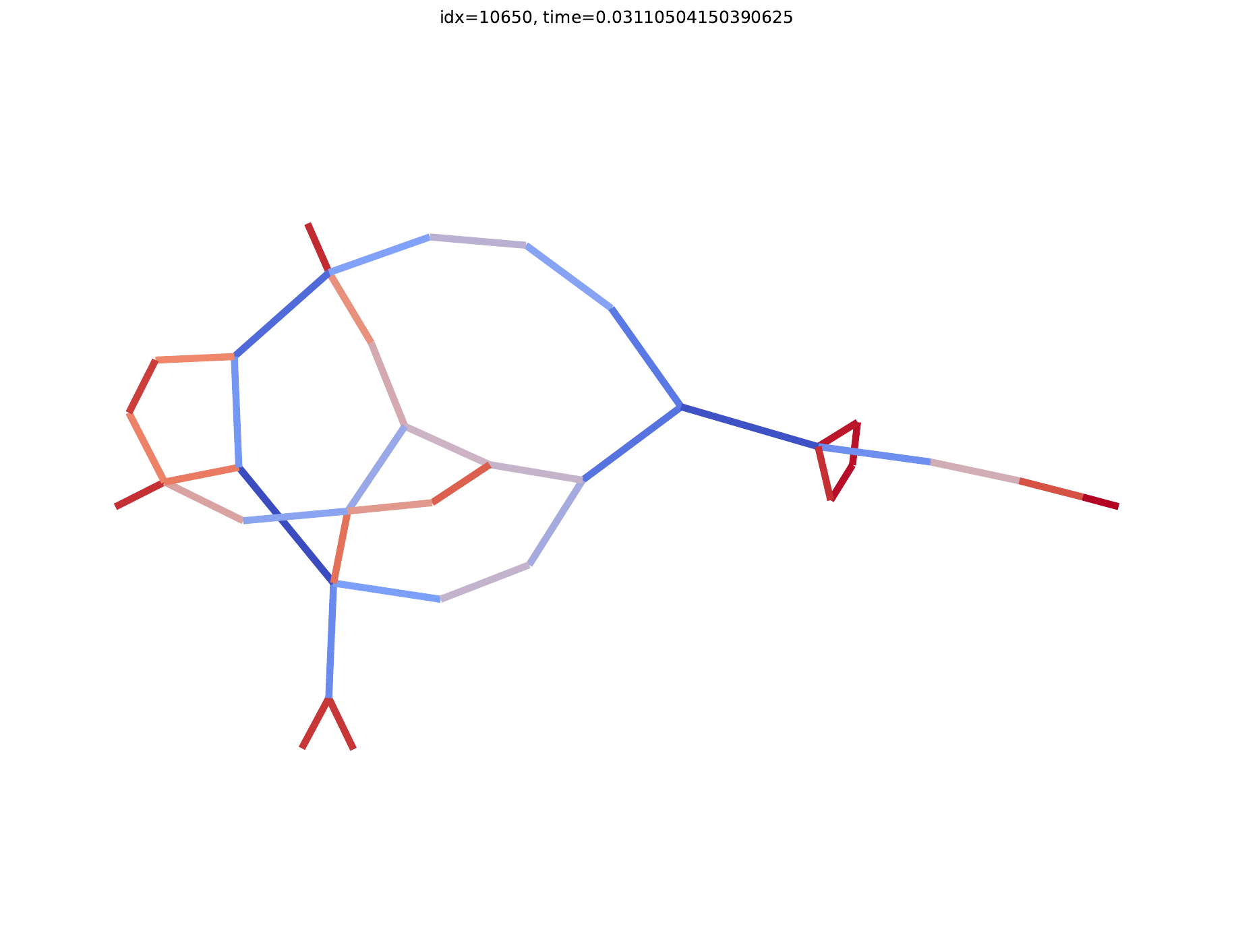} &
\imgcell{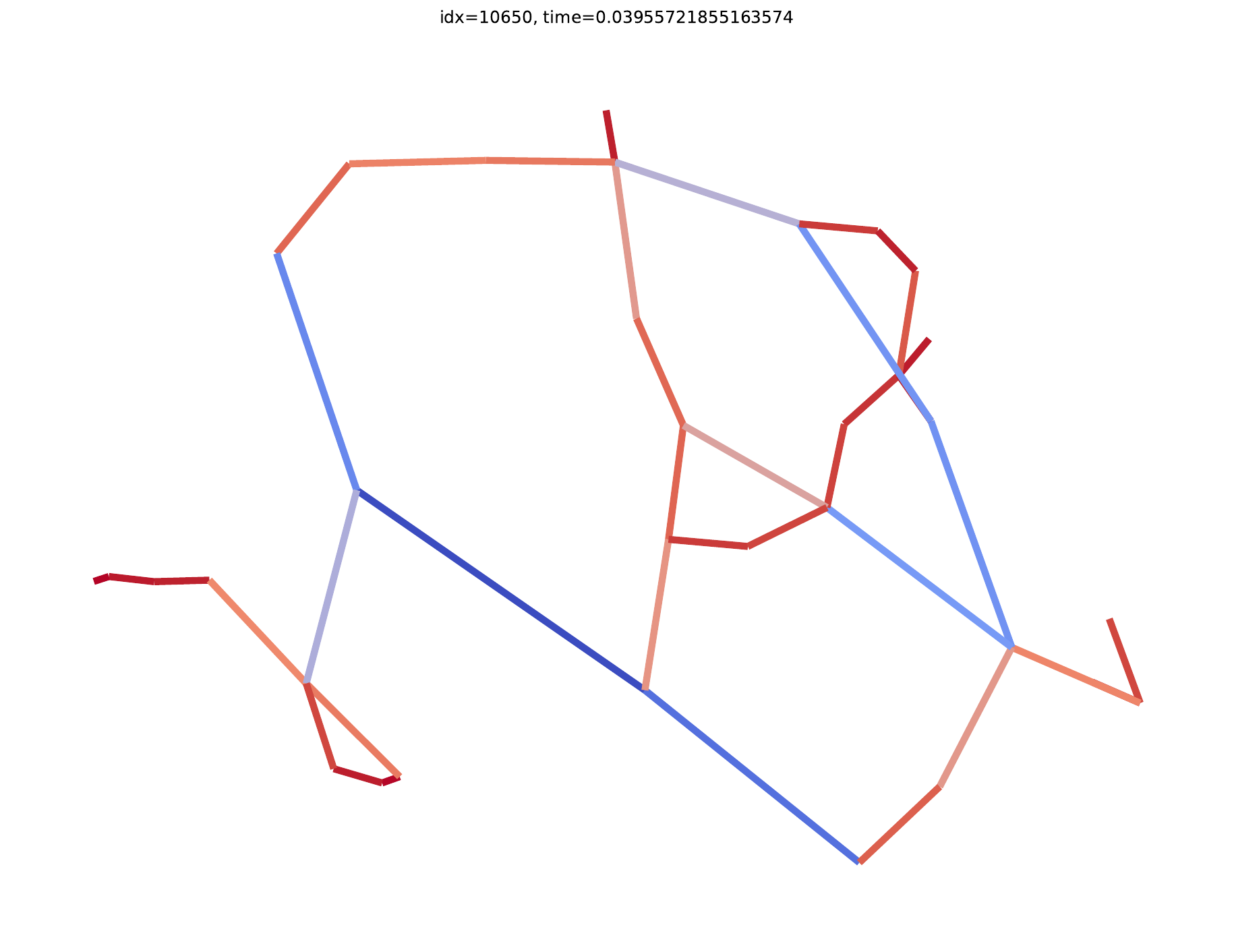} &
\imgcell{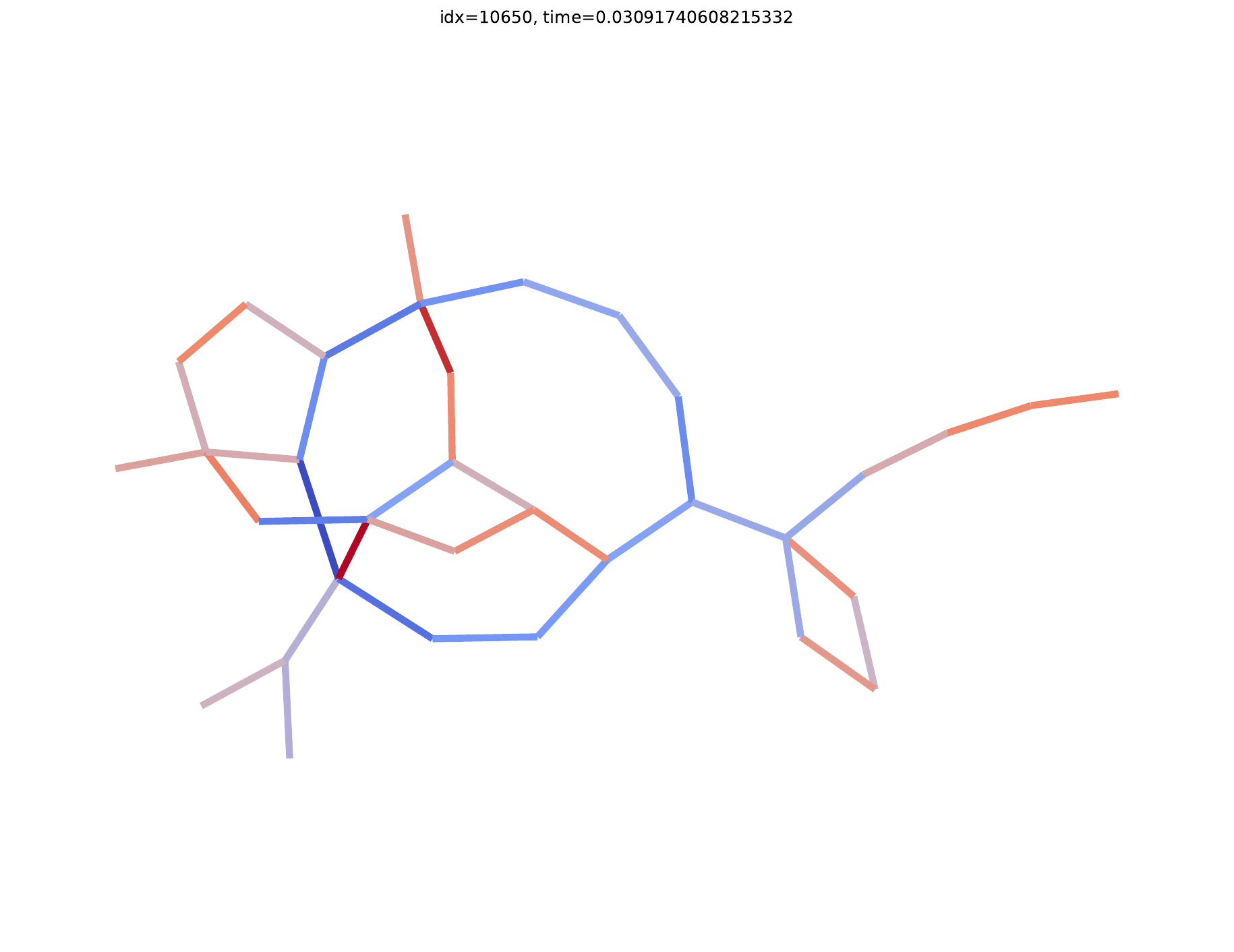} &
\imgcell{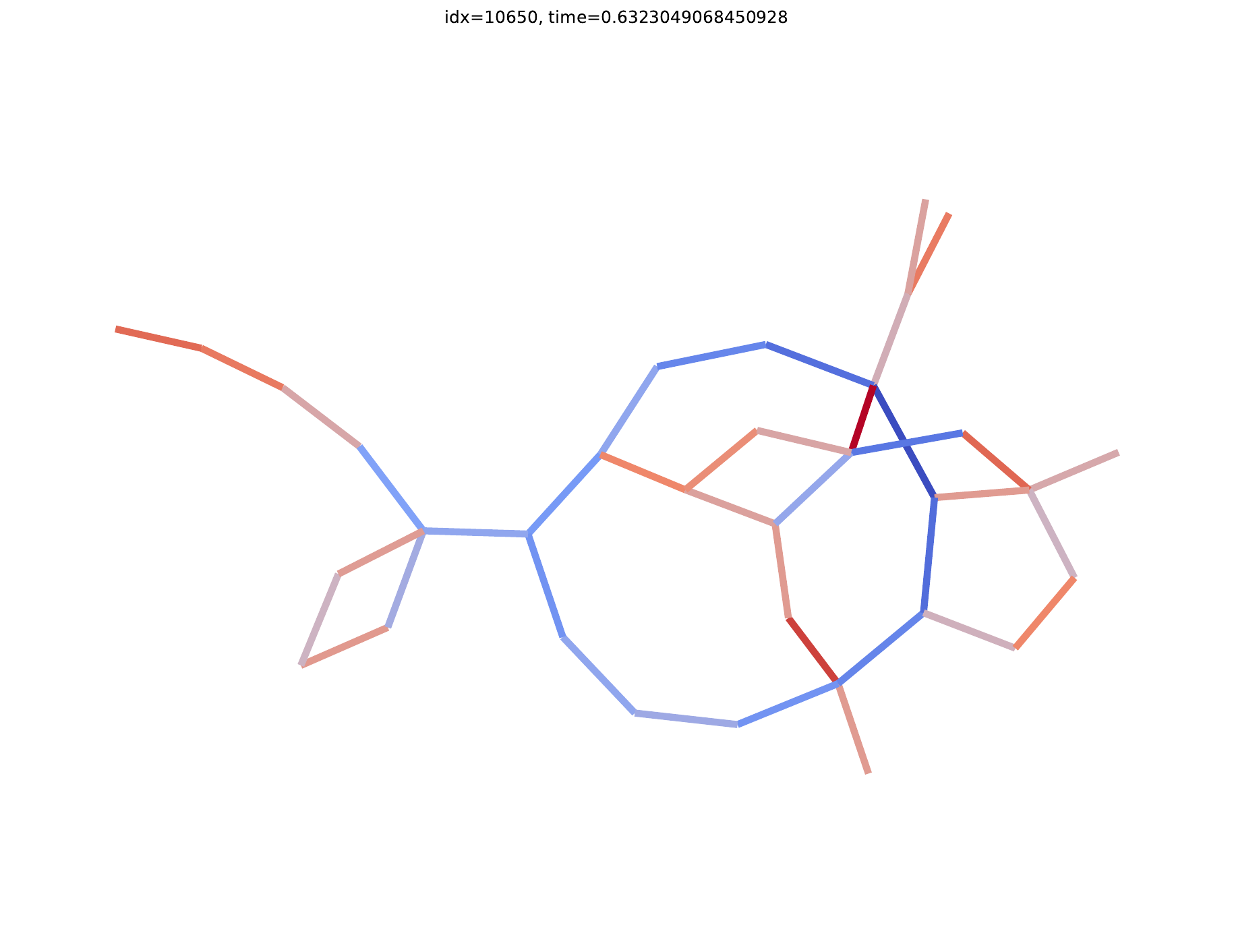} &
\imgcell{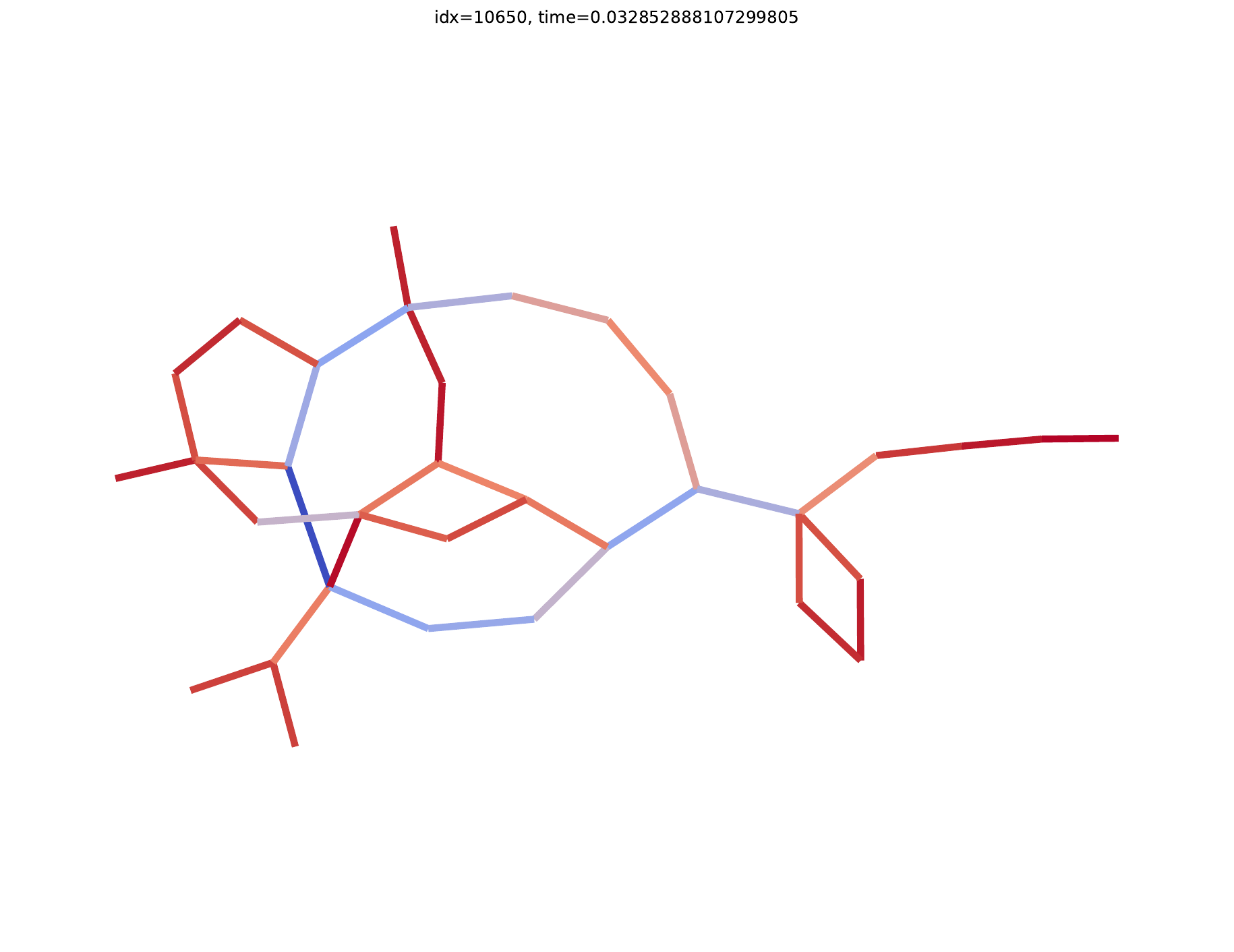} \\

&
t = 0.00s &
t = 0.41s &
t = 0.10s &
t = 0.06s &
t = 98.67s &
t = 0.03s &
t = 0.04s &
t = 0.05s &
t = 0.04s &
t = 0.03s &
t = 0.03s &
t = 0.03s \\

\makecell{\bfseries grafo2300.17\\N = 33\\M = 38} &
\imgcell{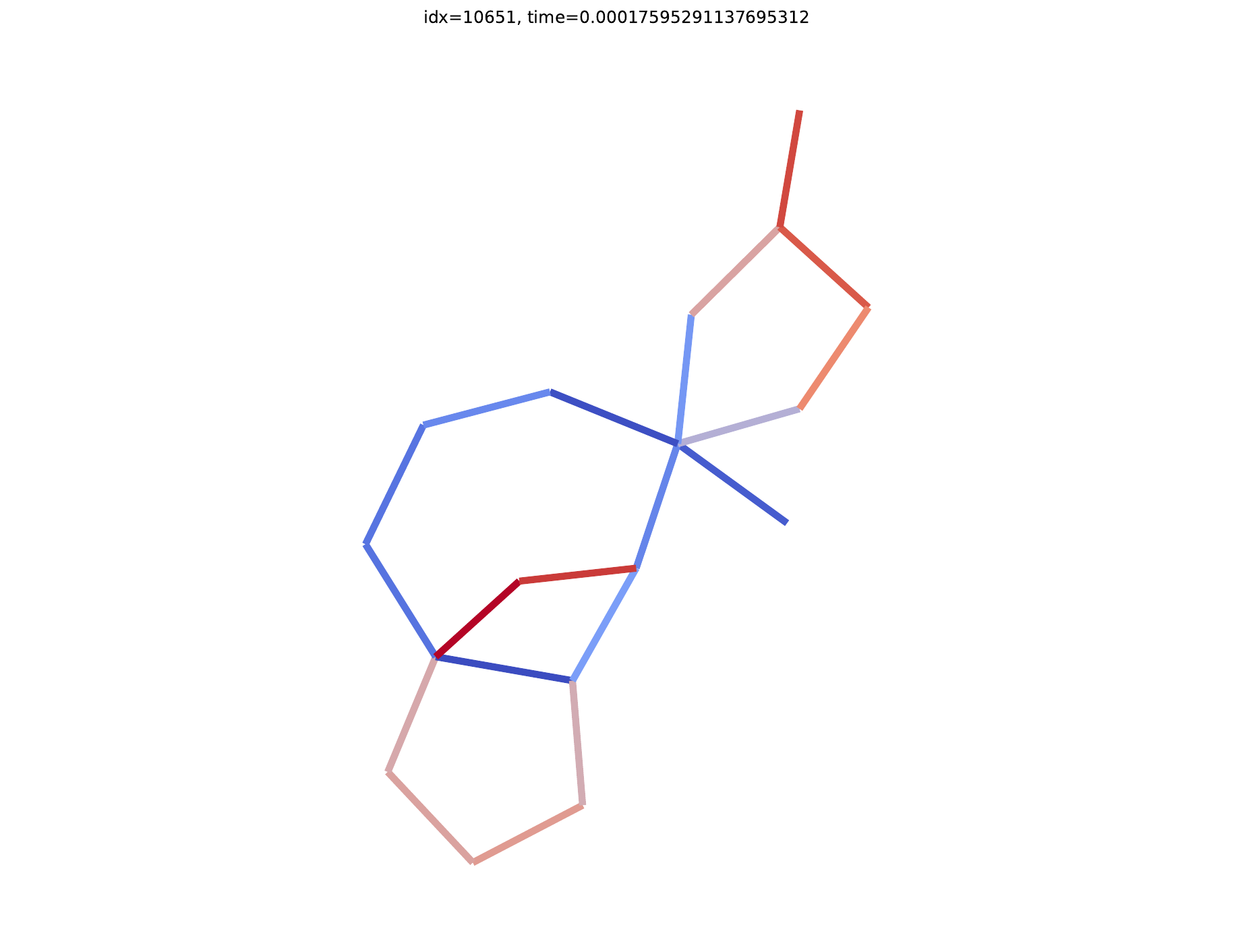} &
\imgcell{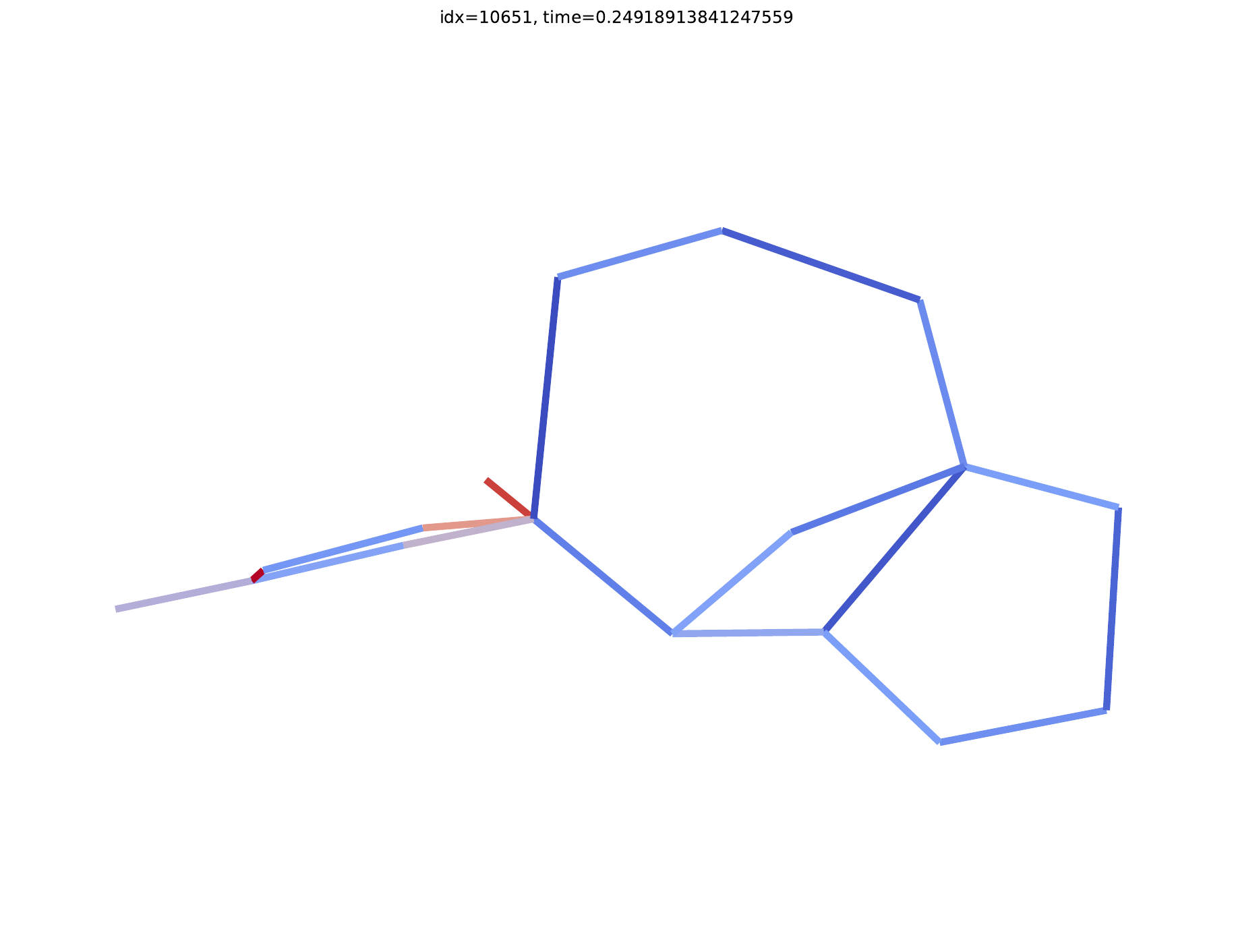} &
\imgcell{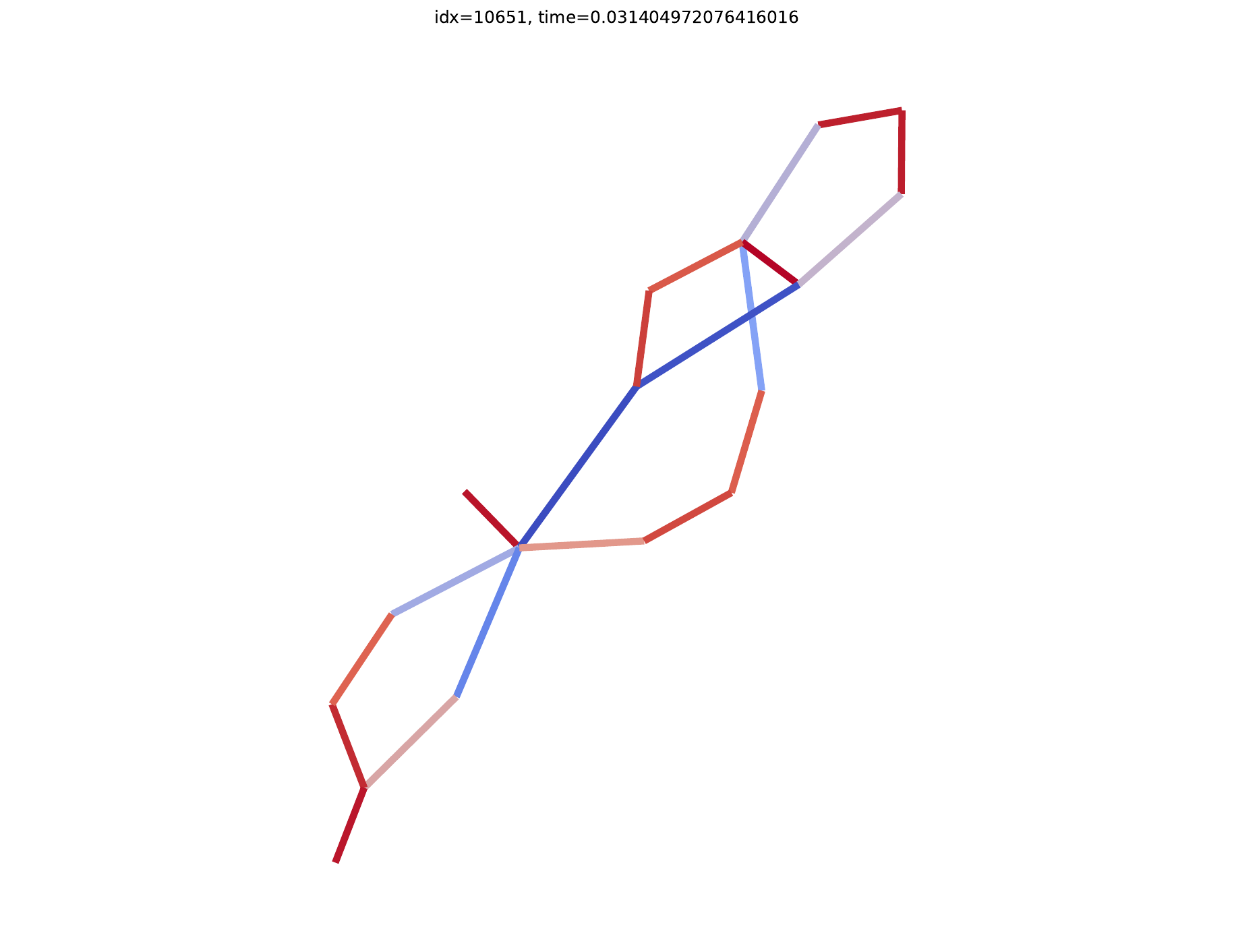} &
\imgcell{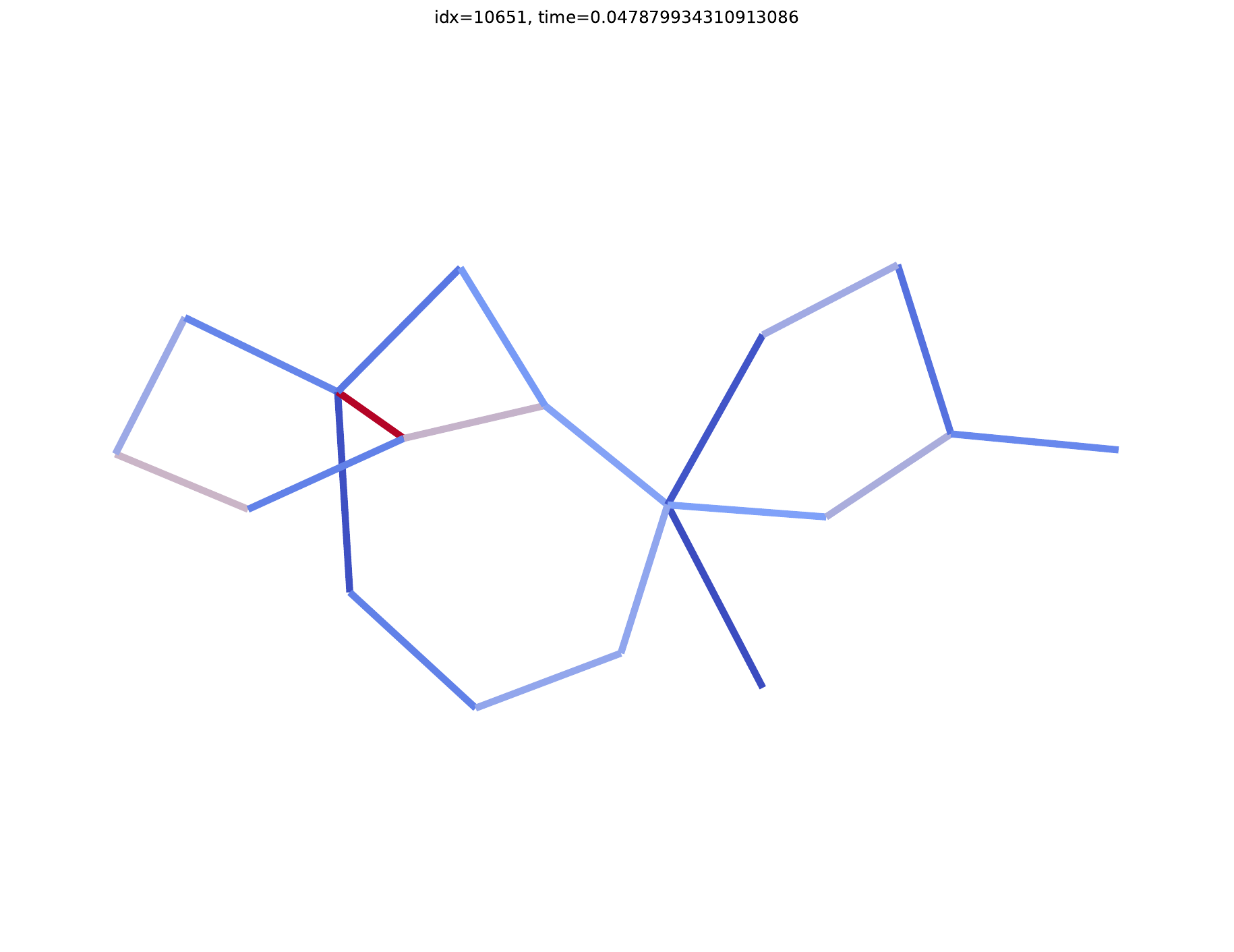} &
\imgcell{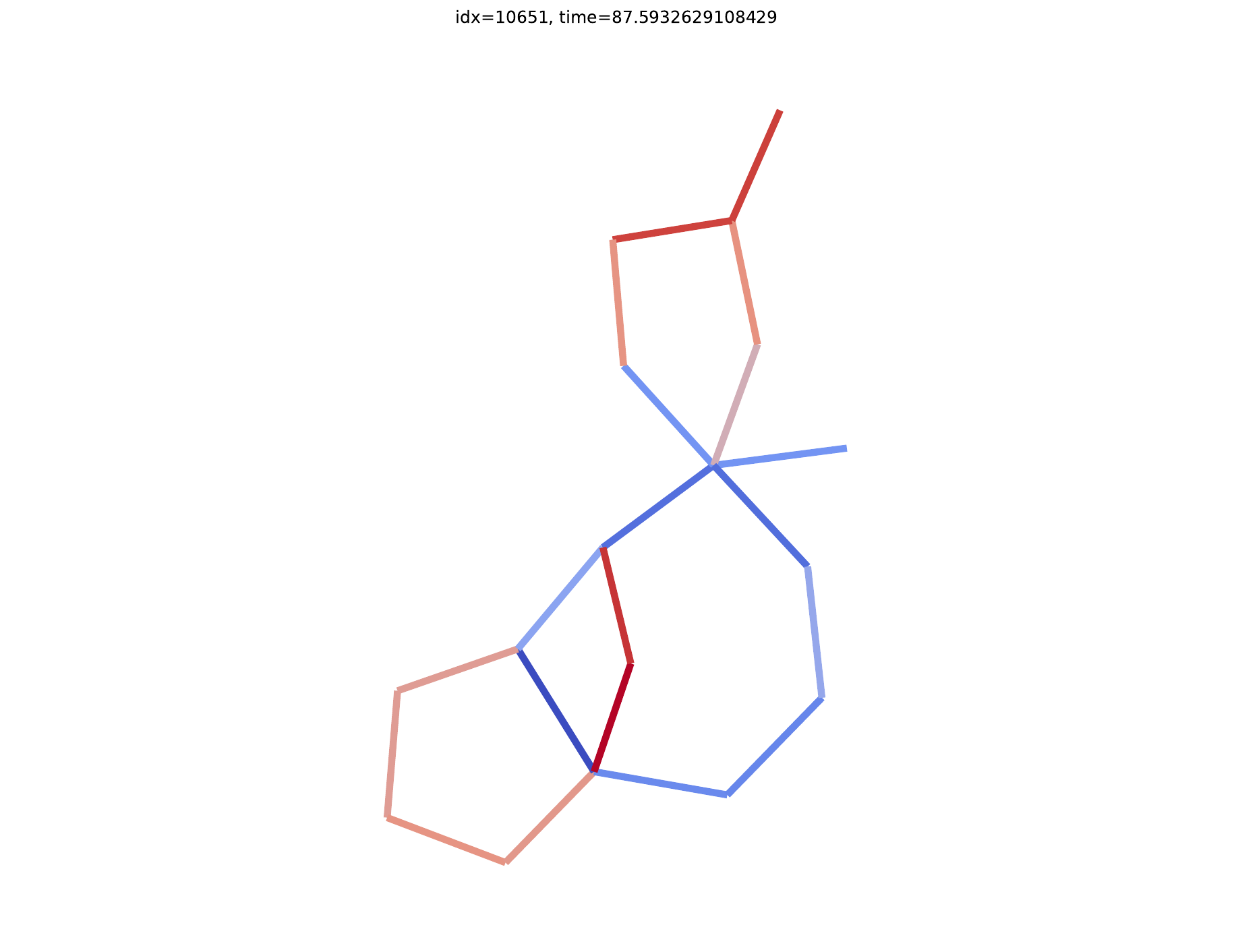} &
\imgcell{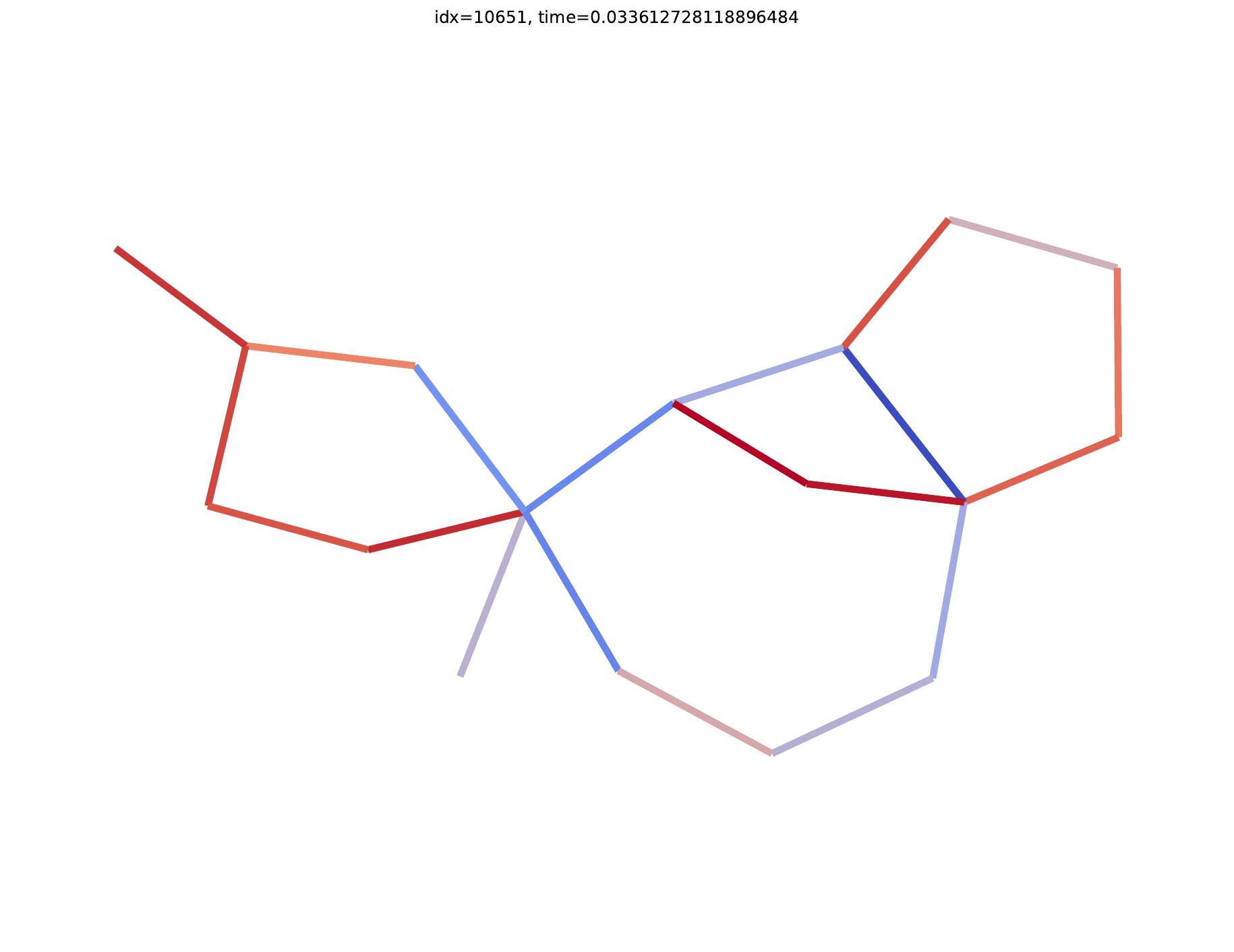} &
\imgcell{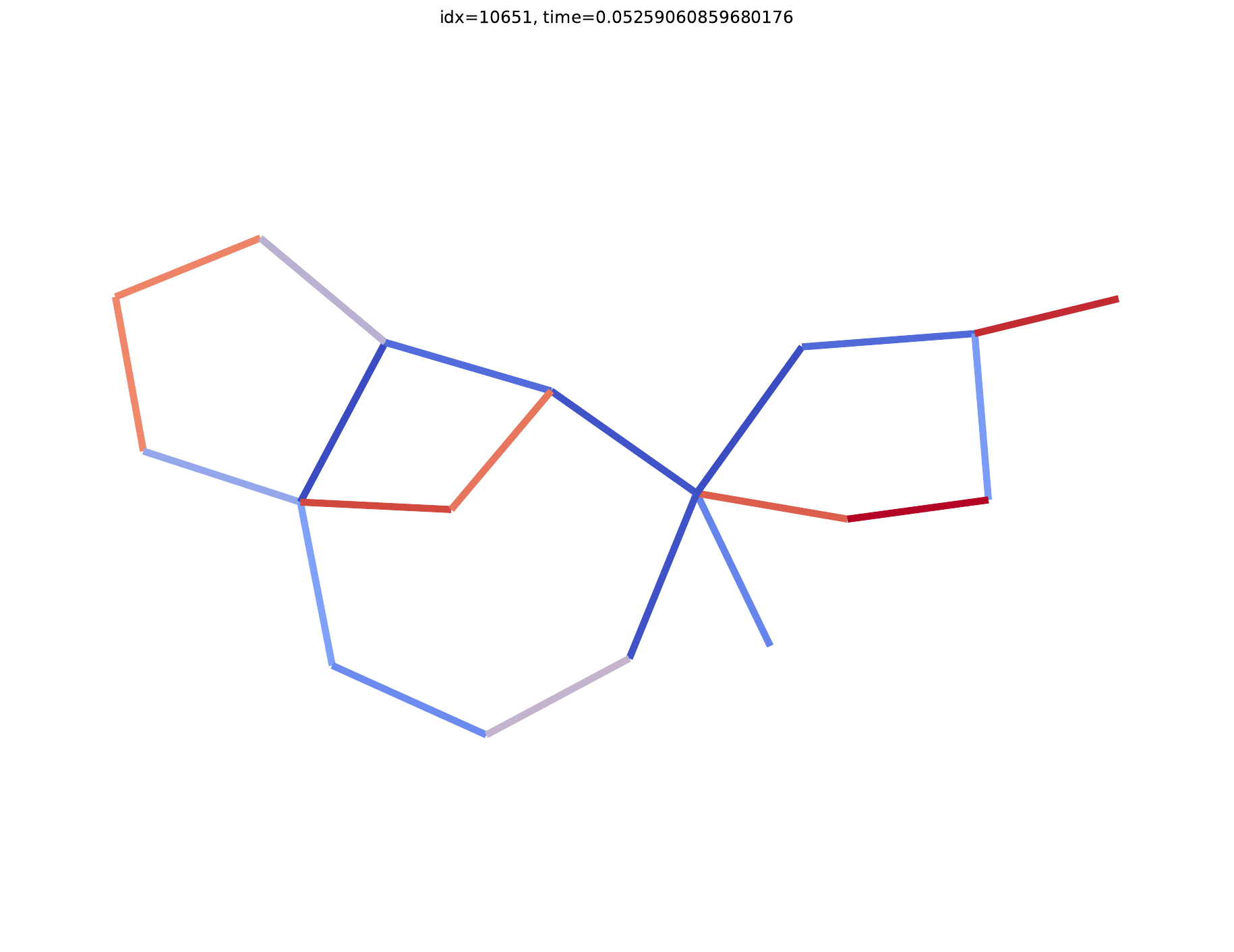} &
\imgcell{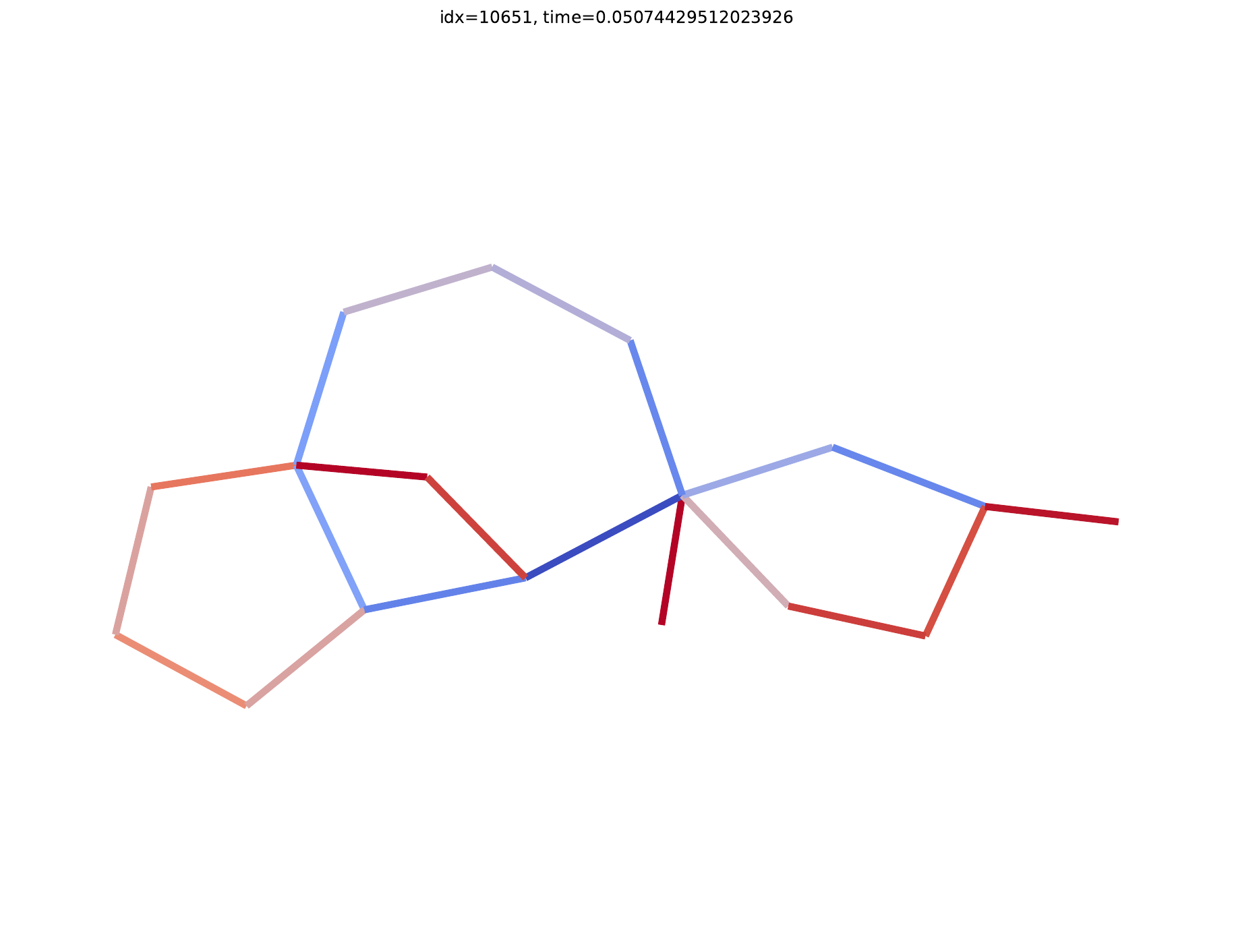} &
\imgcell{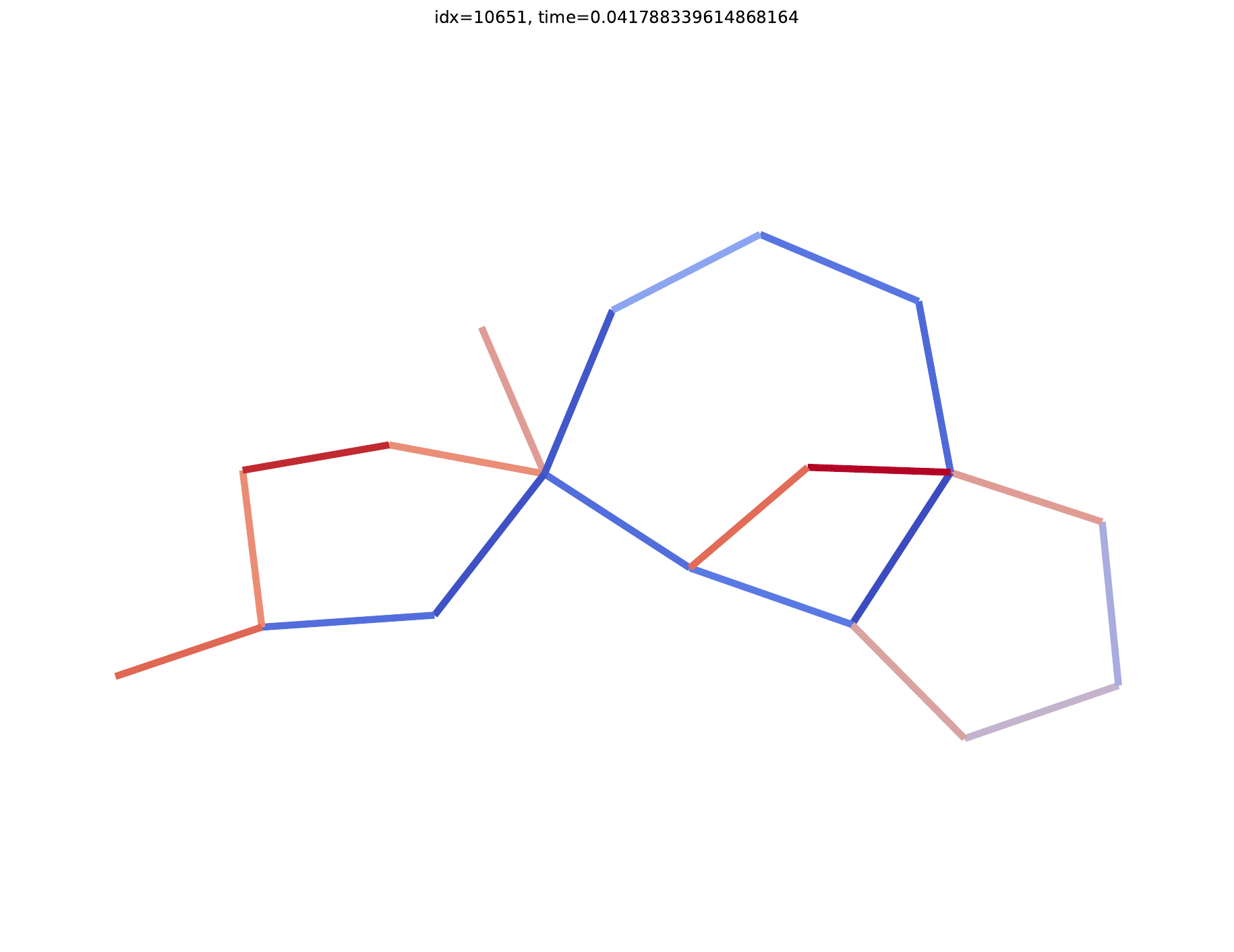} &
\imgcell{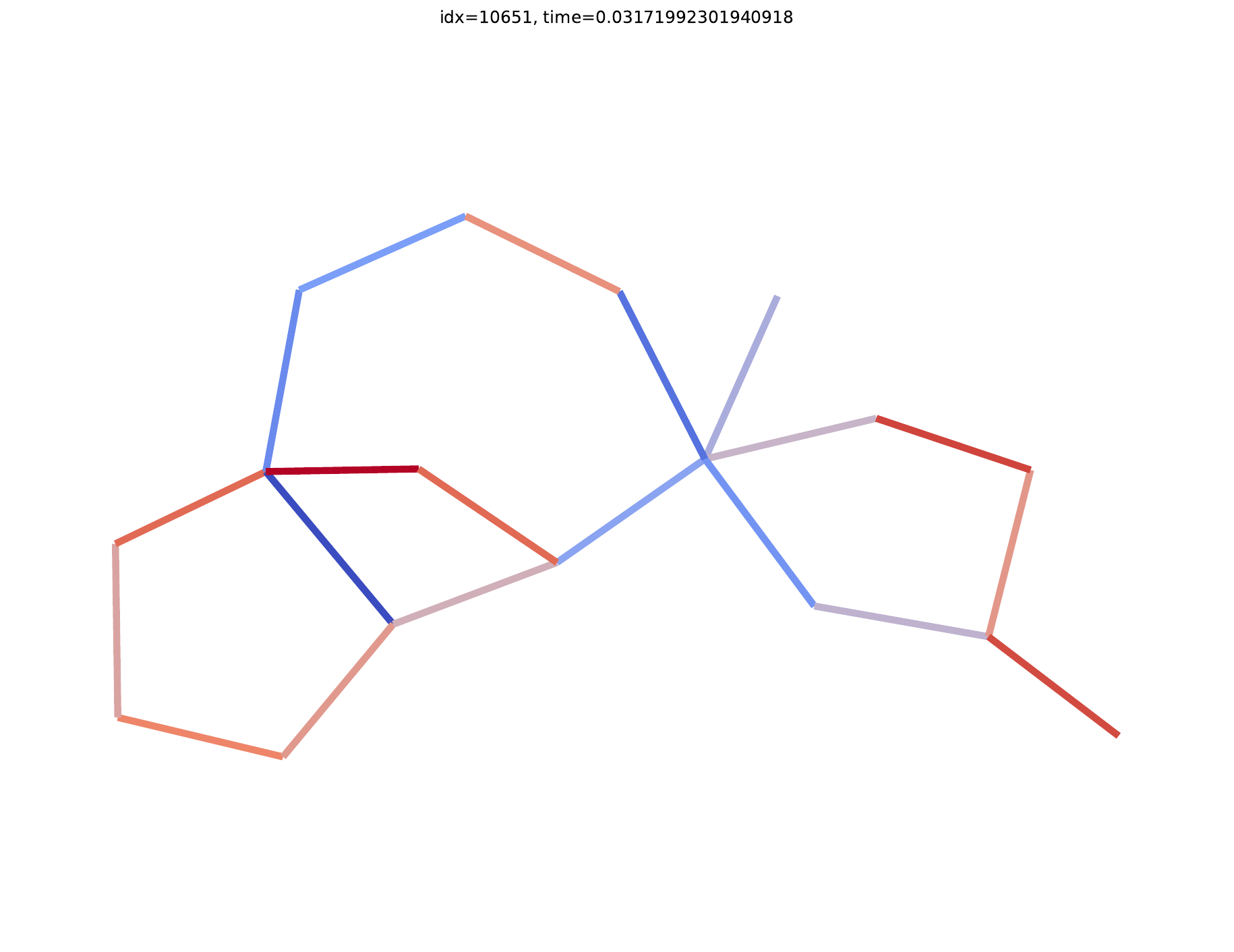} &
\imgcell{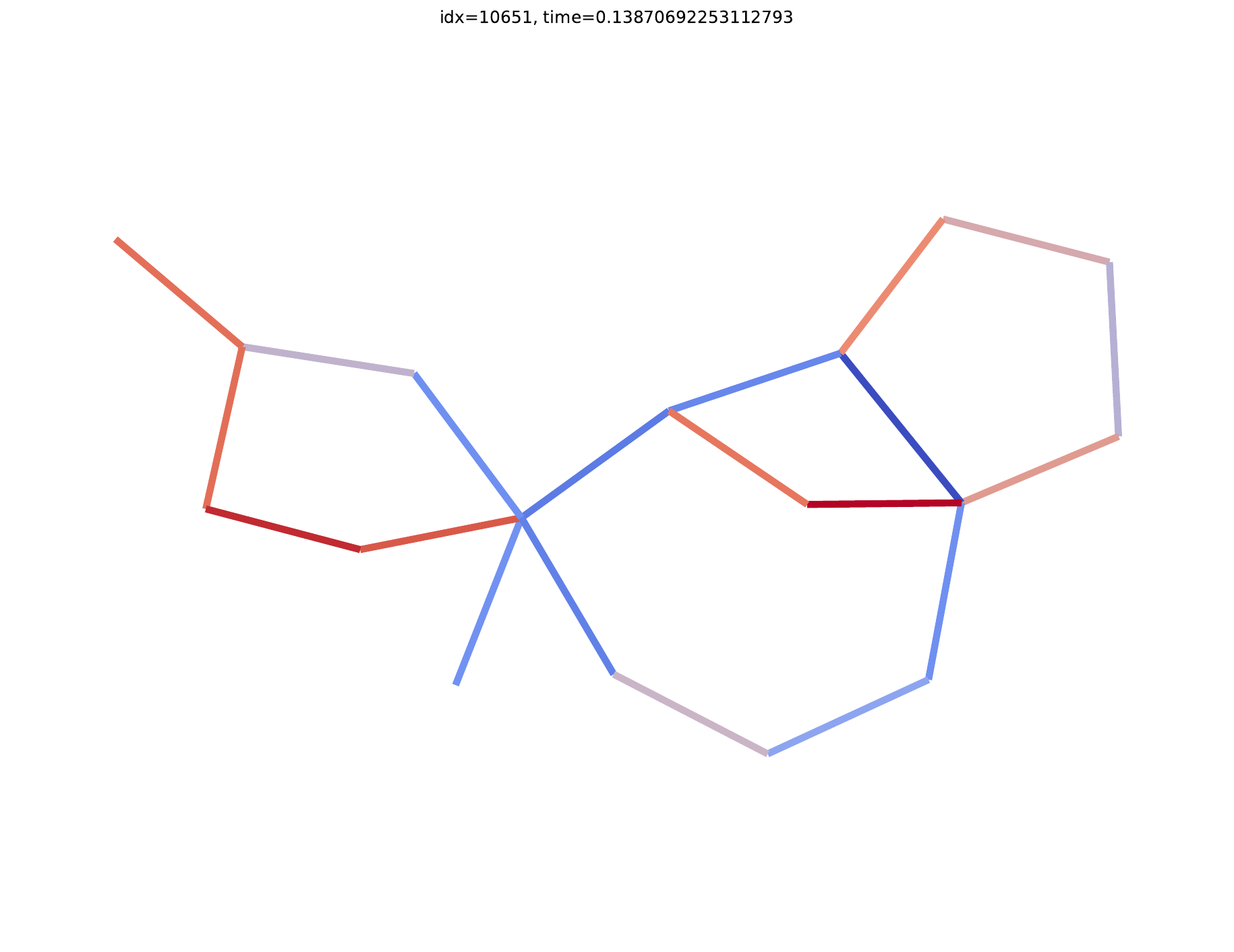} &
\imgcell{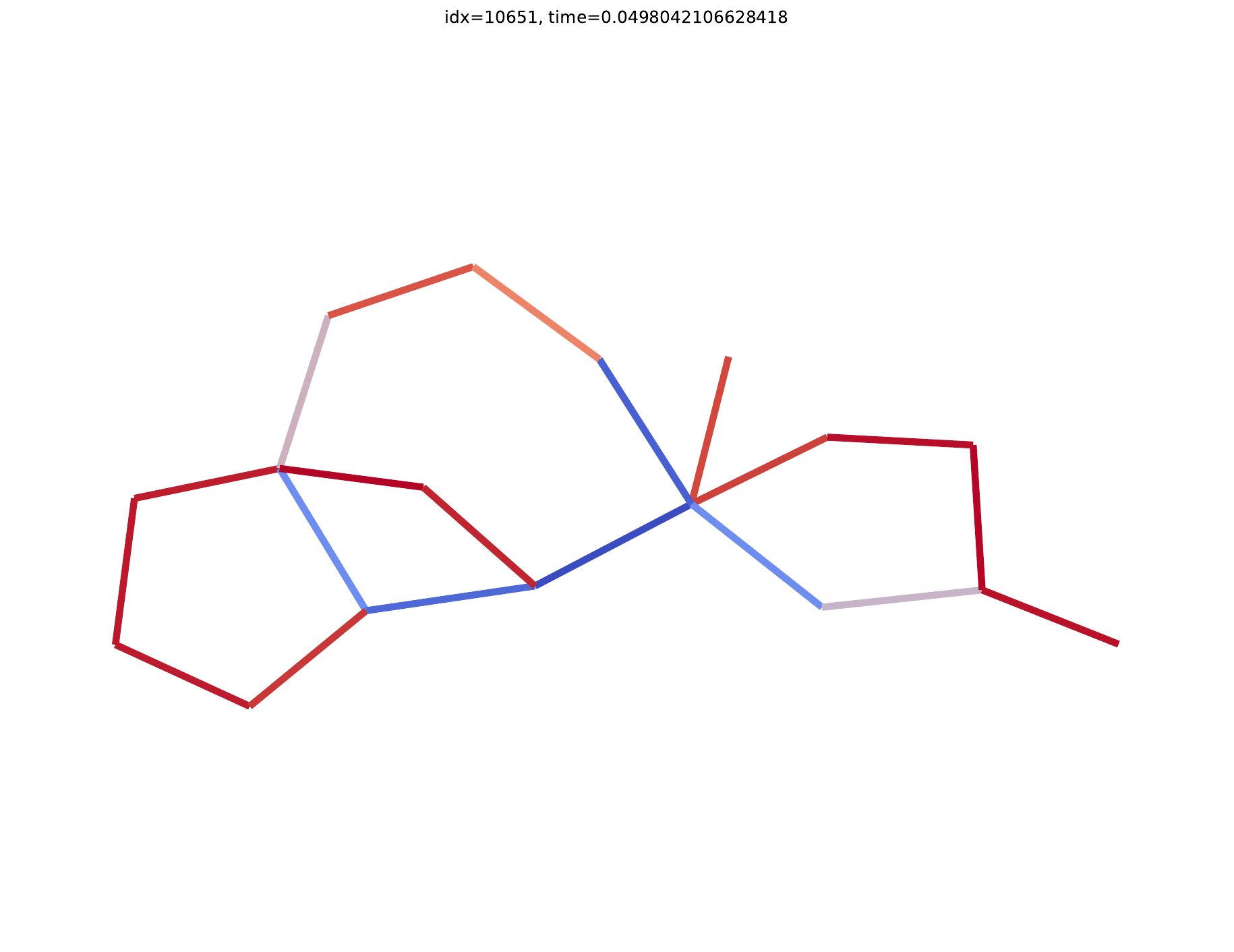} \\

&
t = 0.00s &
t = 0.25s &
t = 0.03s &
t = 0.05s &
t = 87.59s &
t = 0.03s &
t = 0.05s &
t = 0.04s &
t = 0.04s &
t = 0.03s &
t = 0.04s &
t = 0.05s \\

\makecell{\bfseries grafo7288.45\\N = 23\\M = 26} &
\imgcell{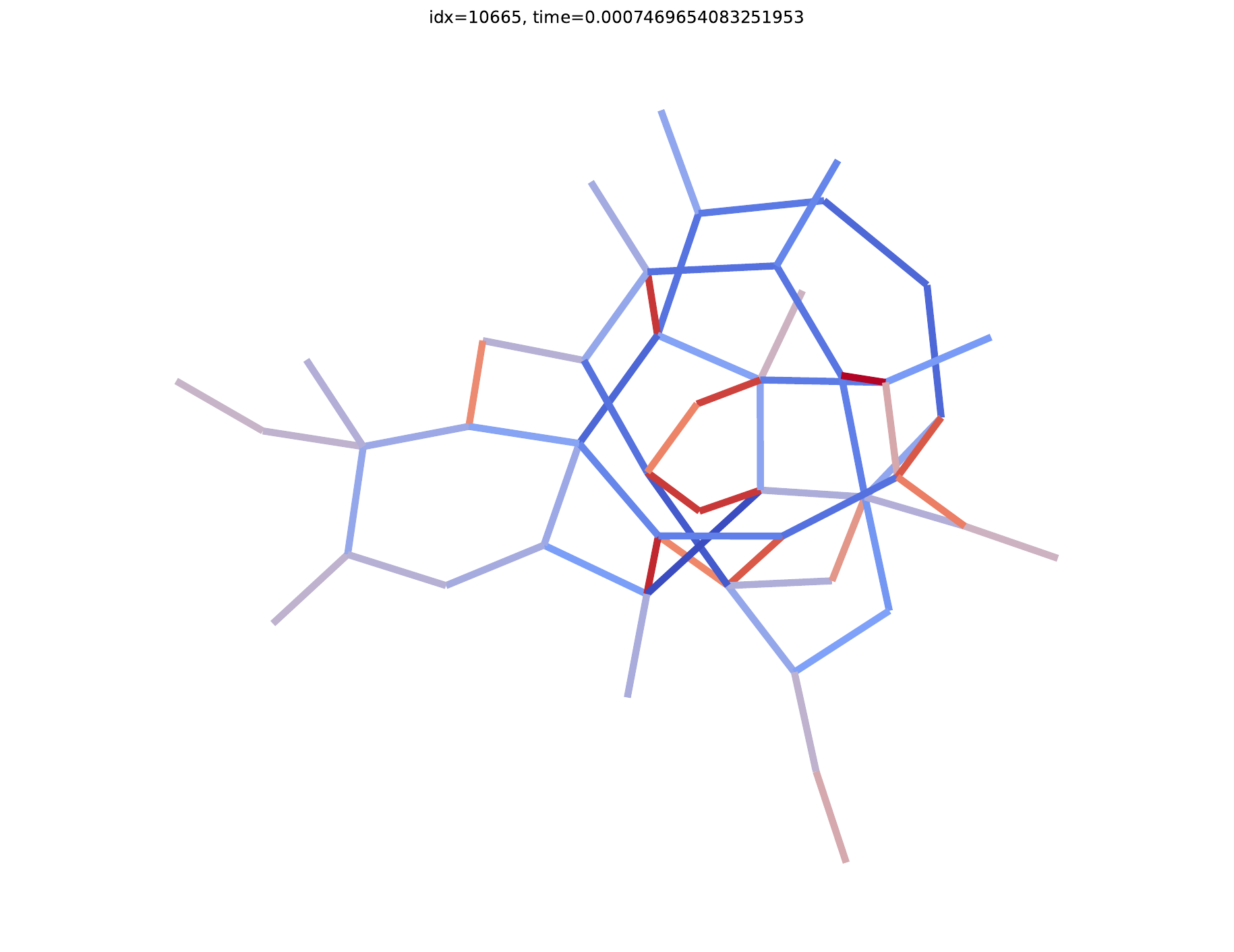} &
\imgcell{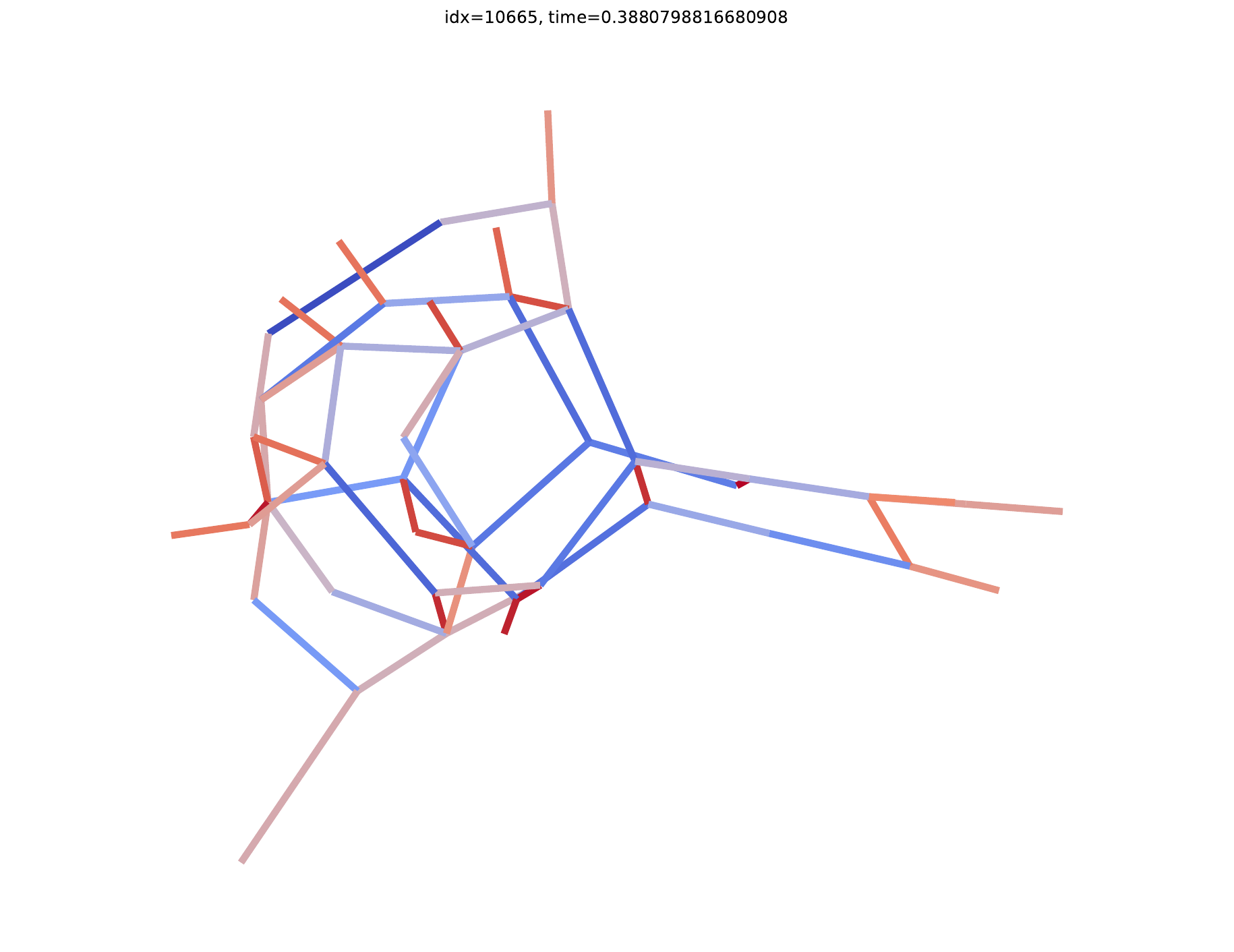} &
\imgcell{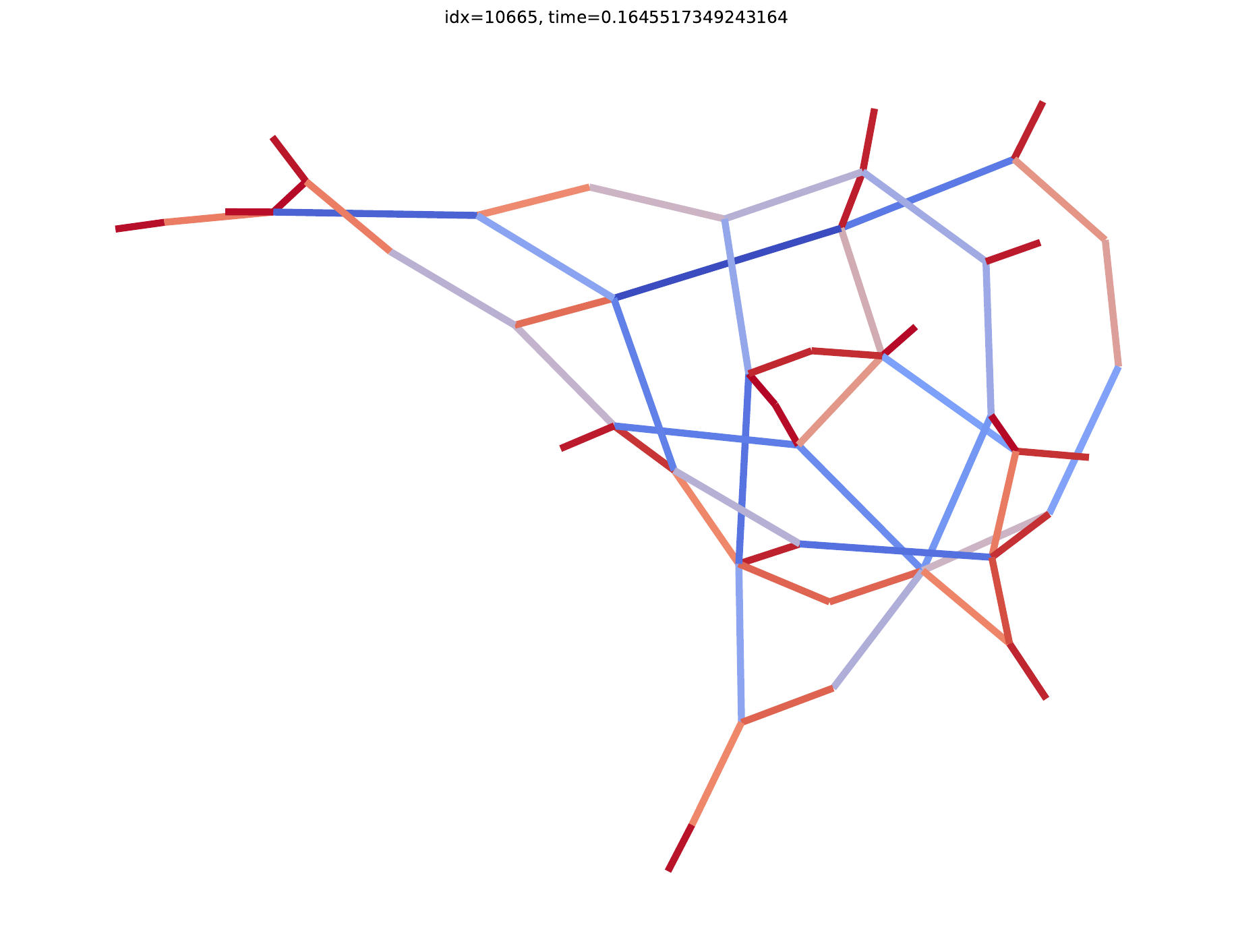} &
\imgcell{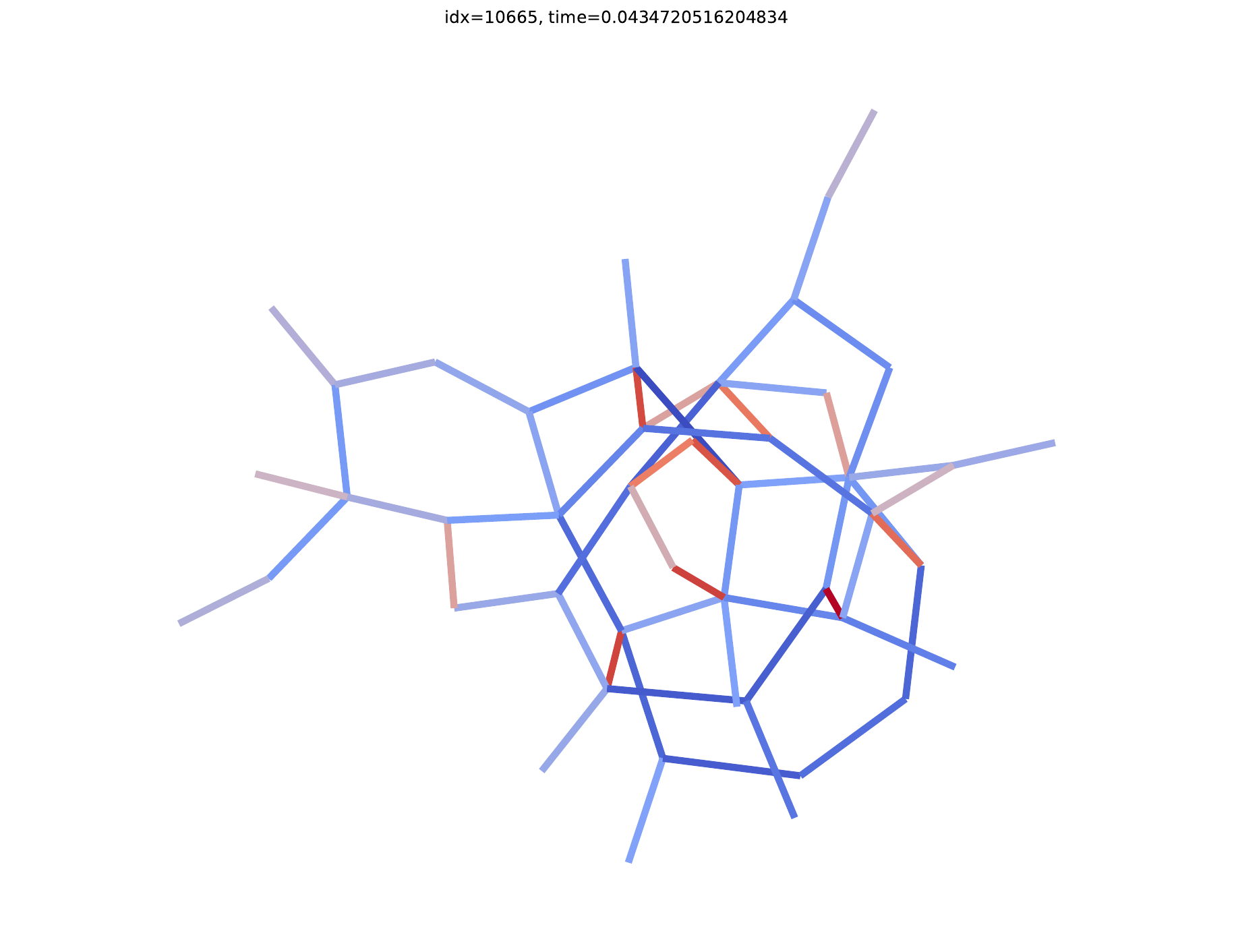} &
\imgcell{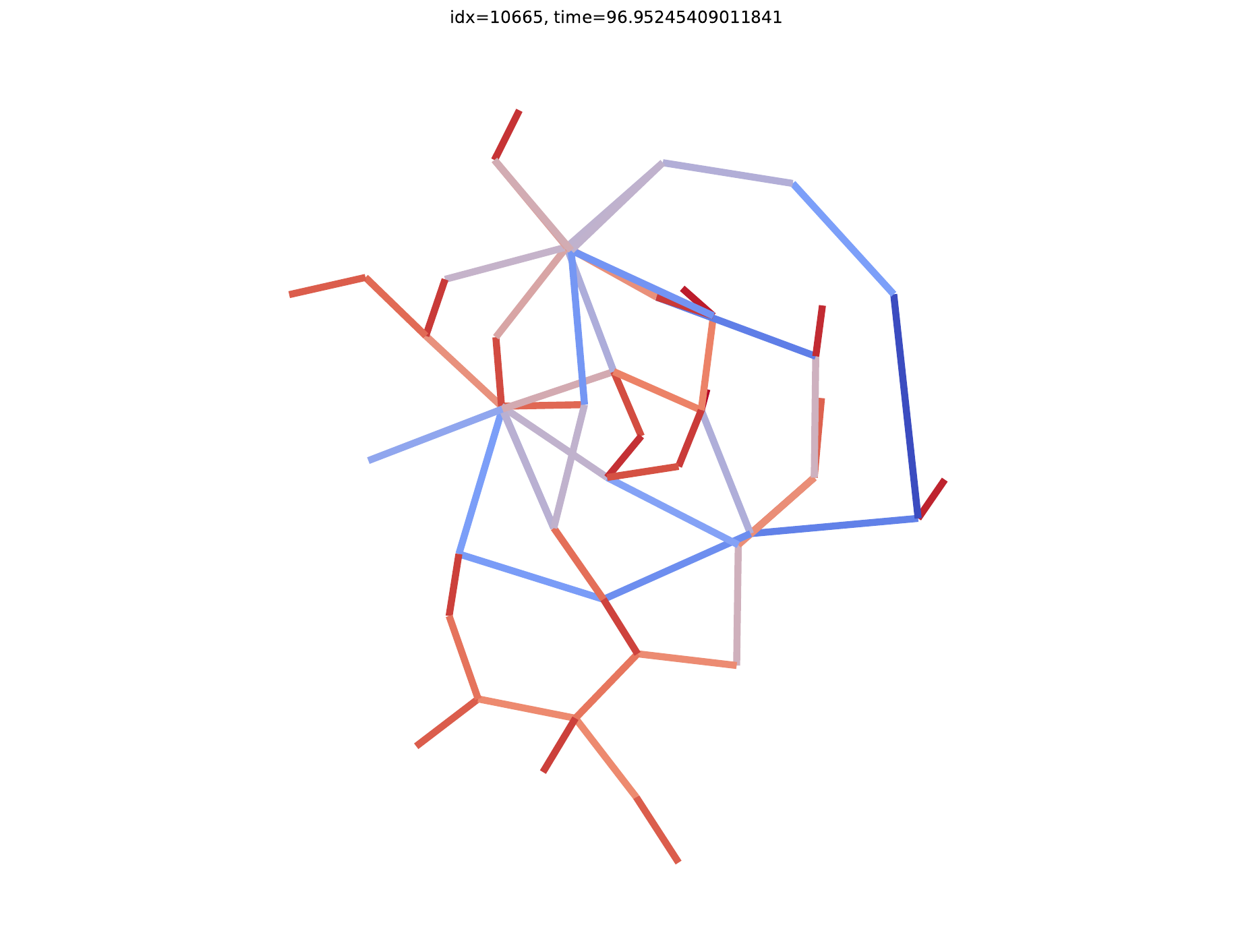} &
\imgcell{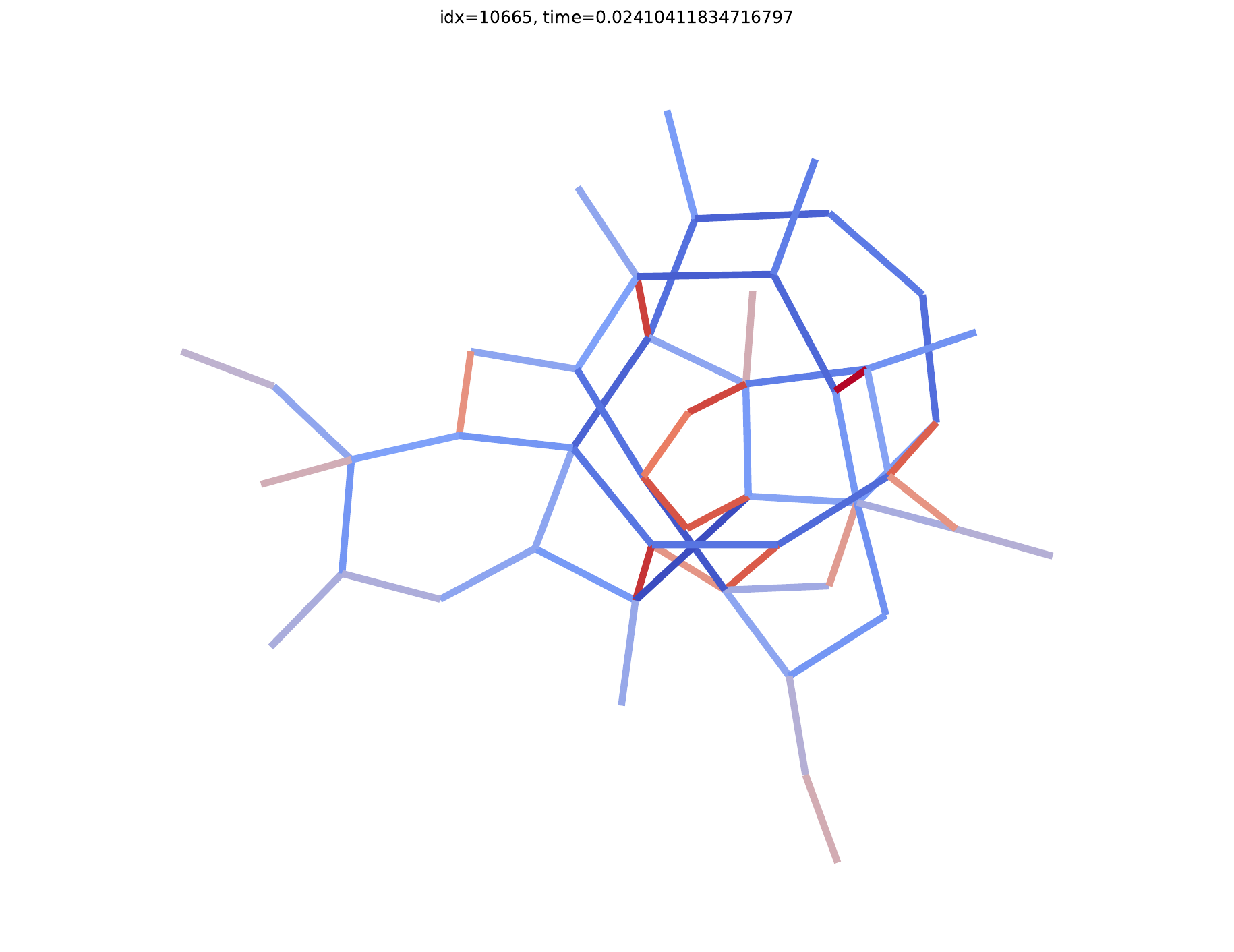} &
\imgcell{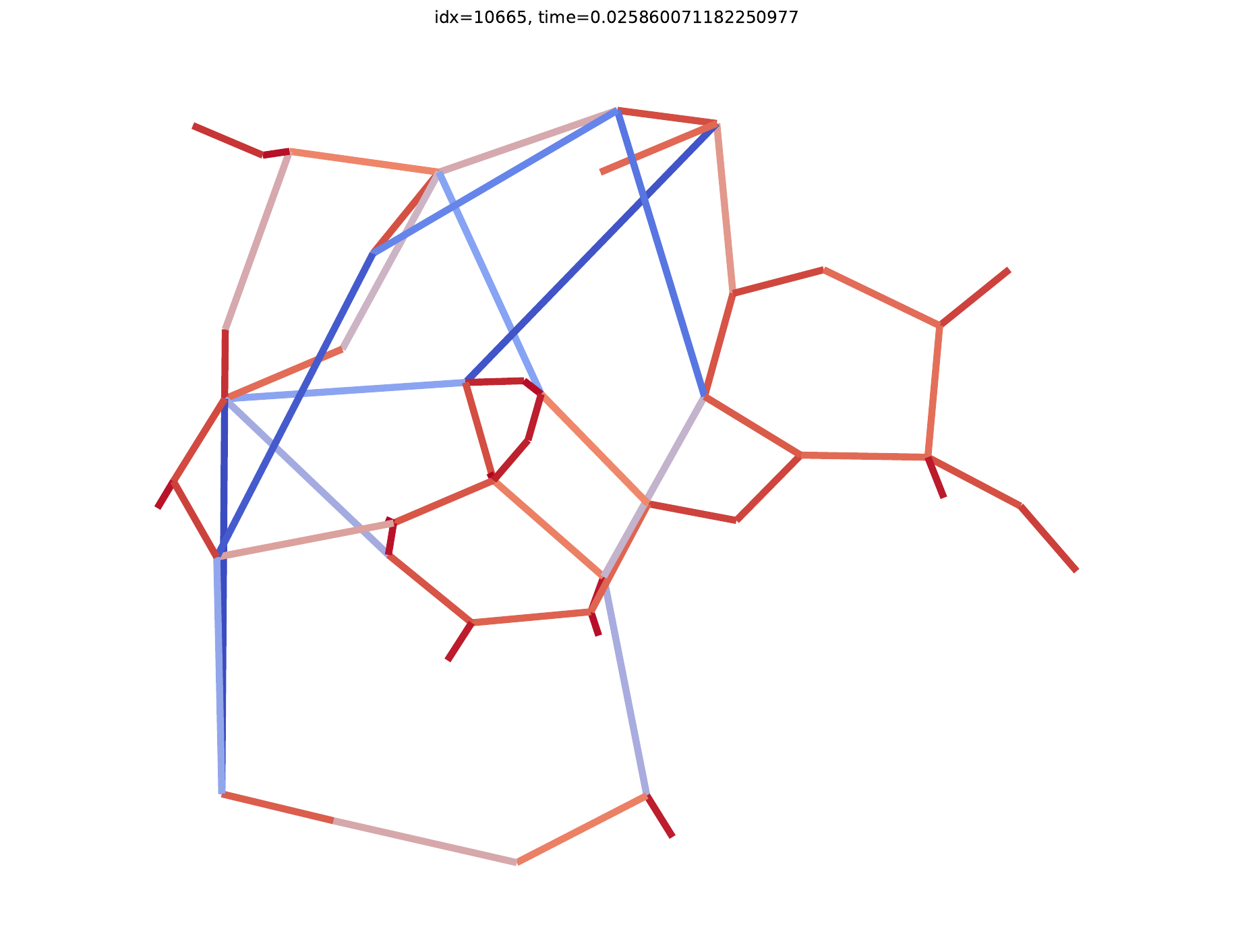} &
\imgcell{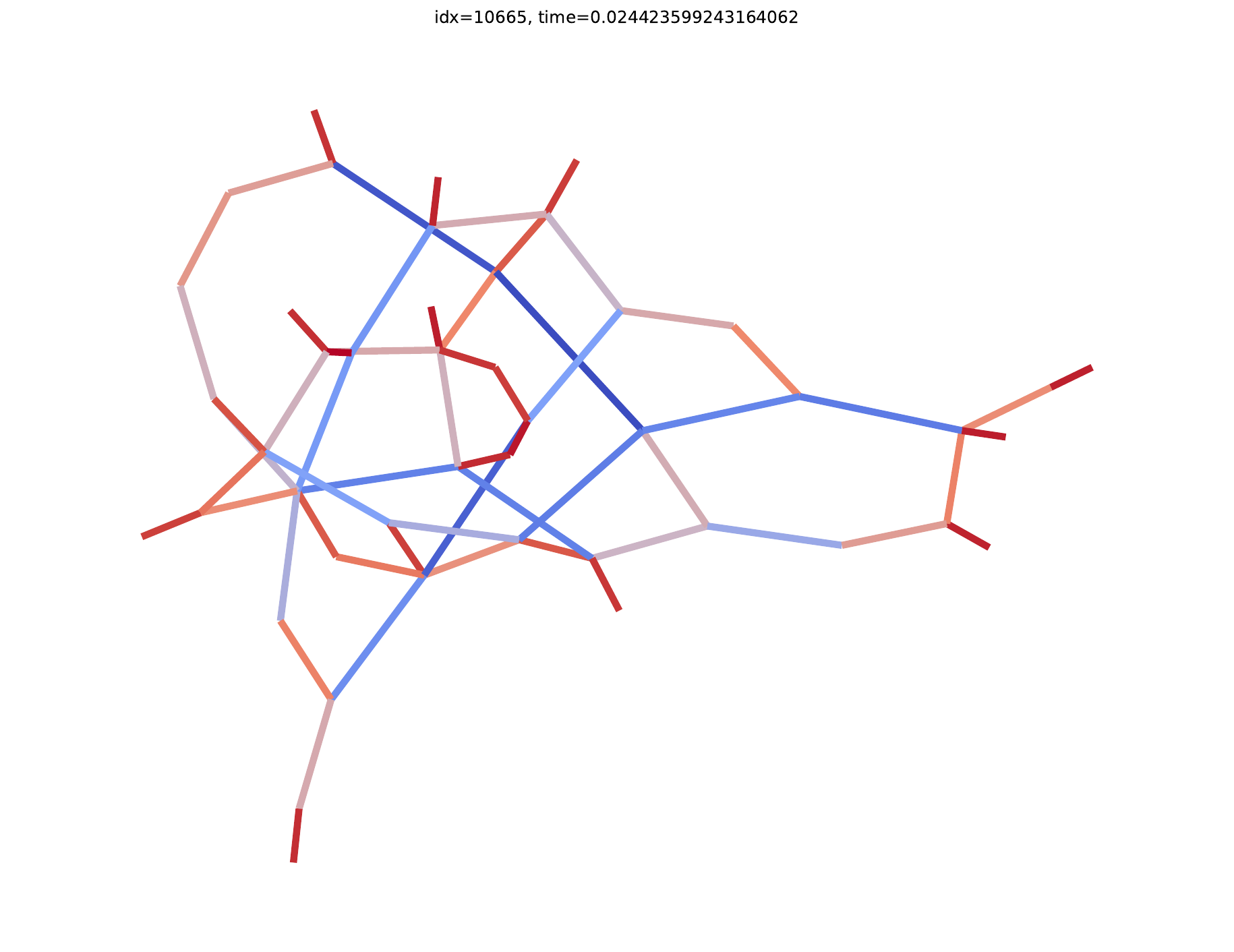} &
\imgcell{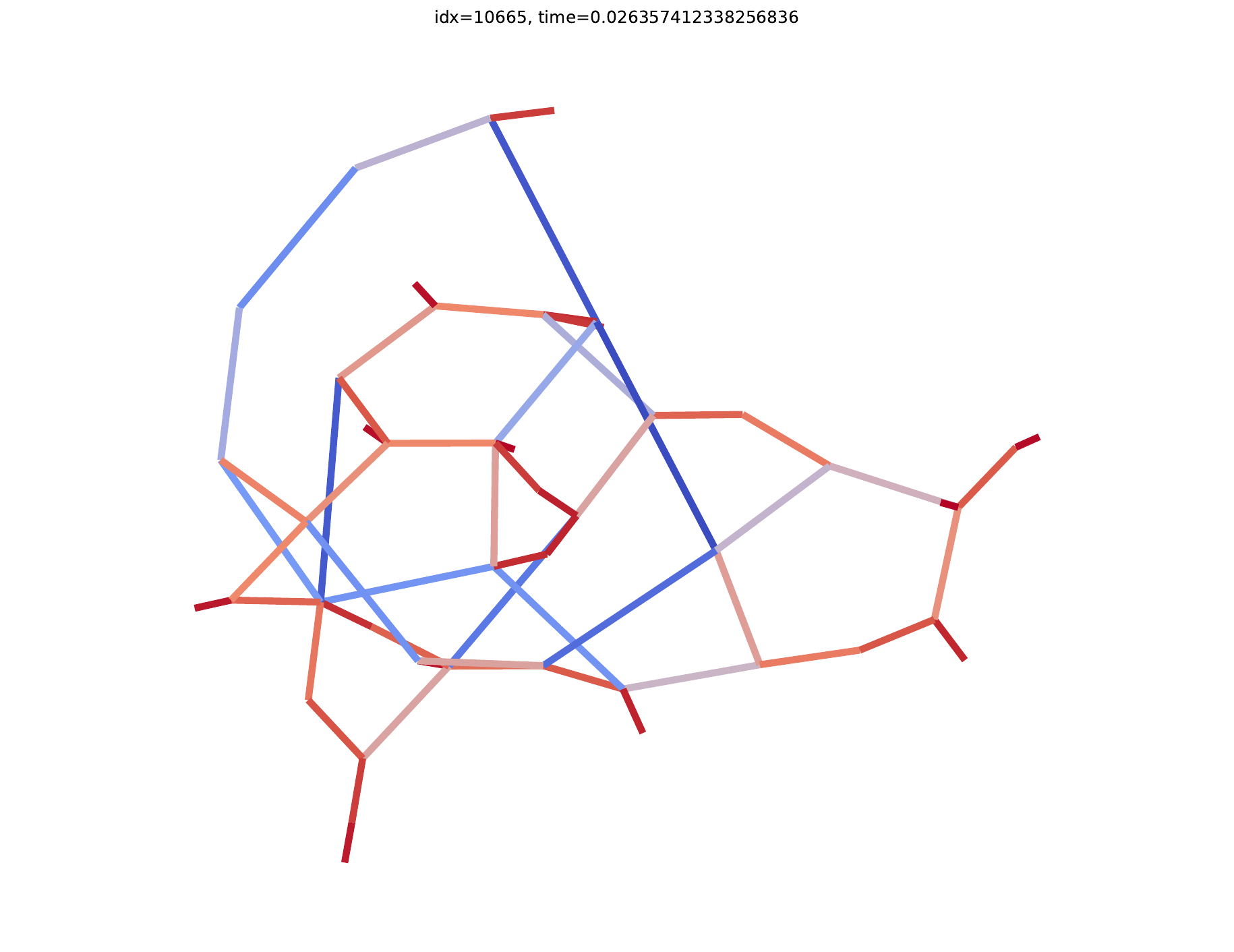} &
\imgcell{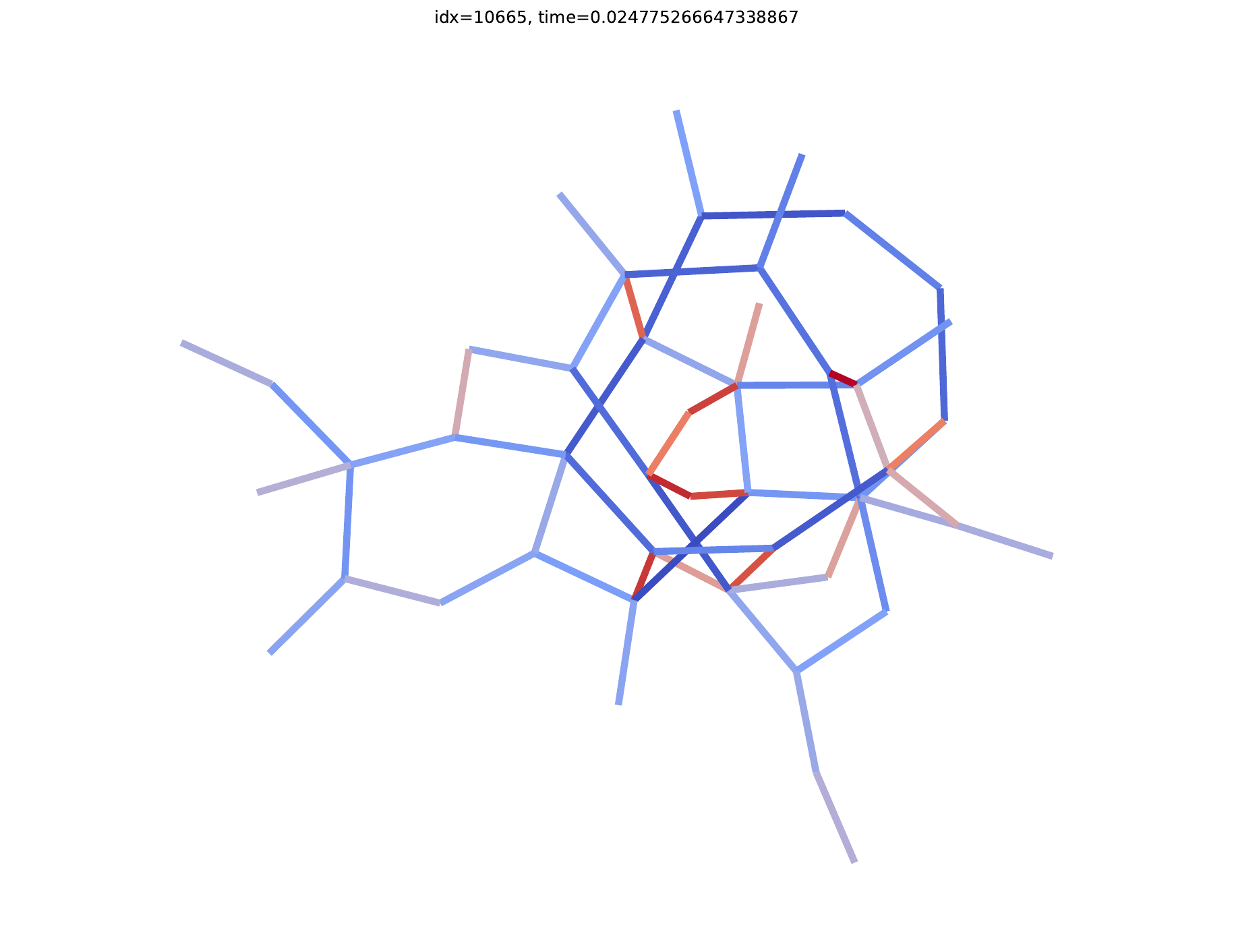} &
\imgcell{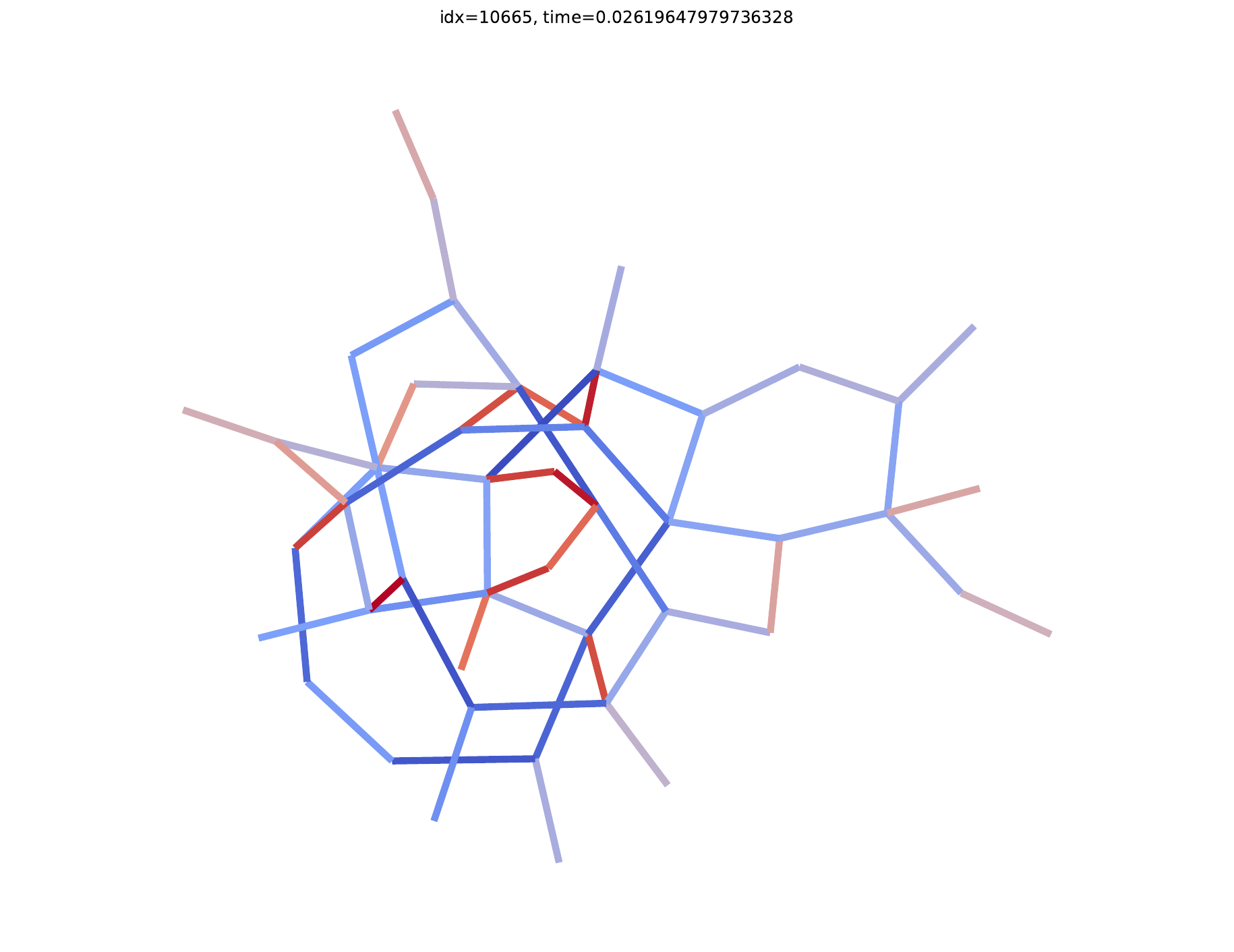} &
\imgcell{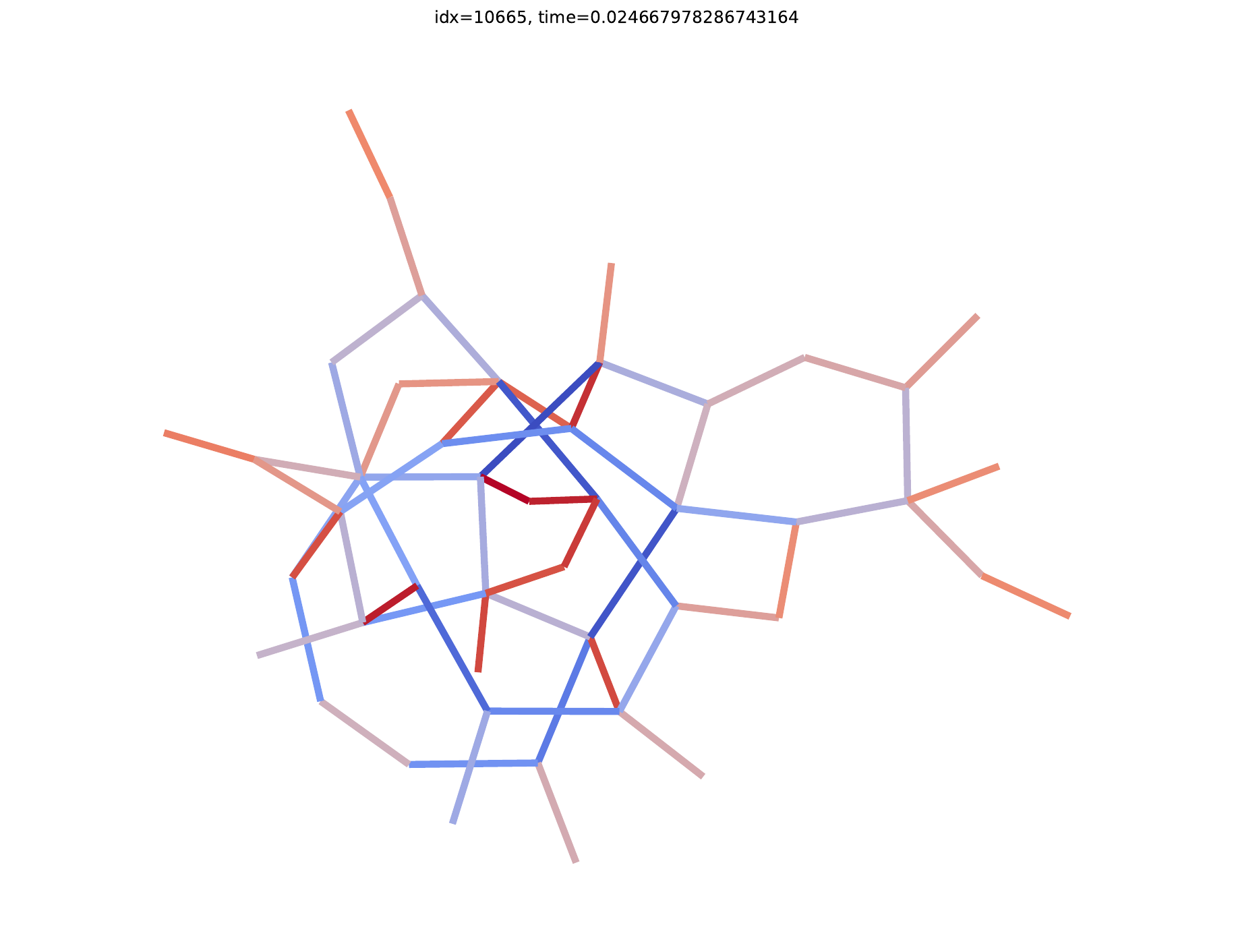} \\

&
t = 0.00s &
t = 0.39s &
t = 0.16s &
t = 0.04s &
t = 96.95s &
t = 0.02s &
t = 0.03s &
t = 0.04s &
t = 0.03s &
t = 0.02s &
t = 0.03s &
t = 0.02s \\

\makecell{\bfseries grafo10343.94\\N = 36\\M = 43} &
\imgcell{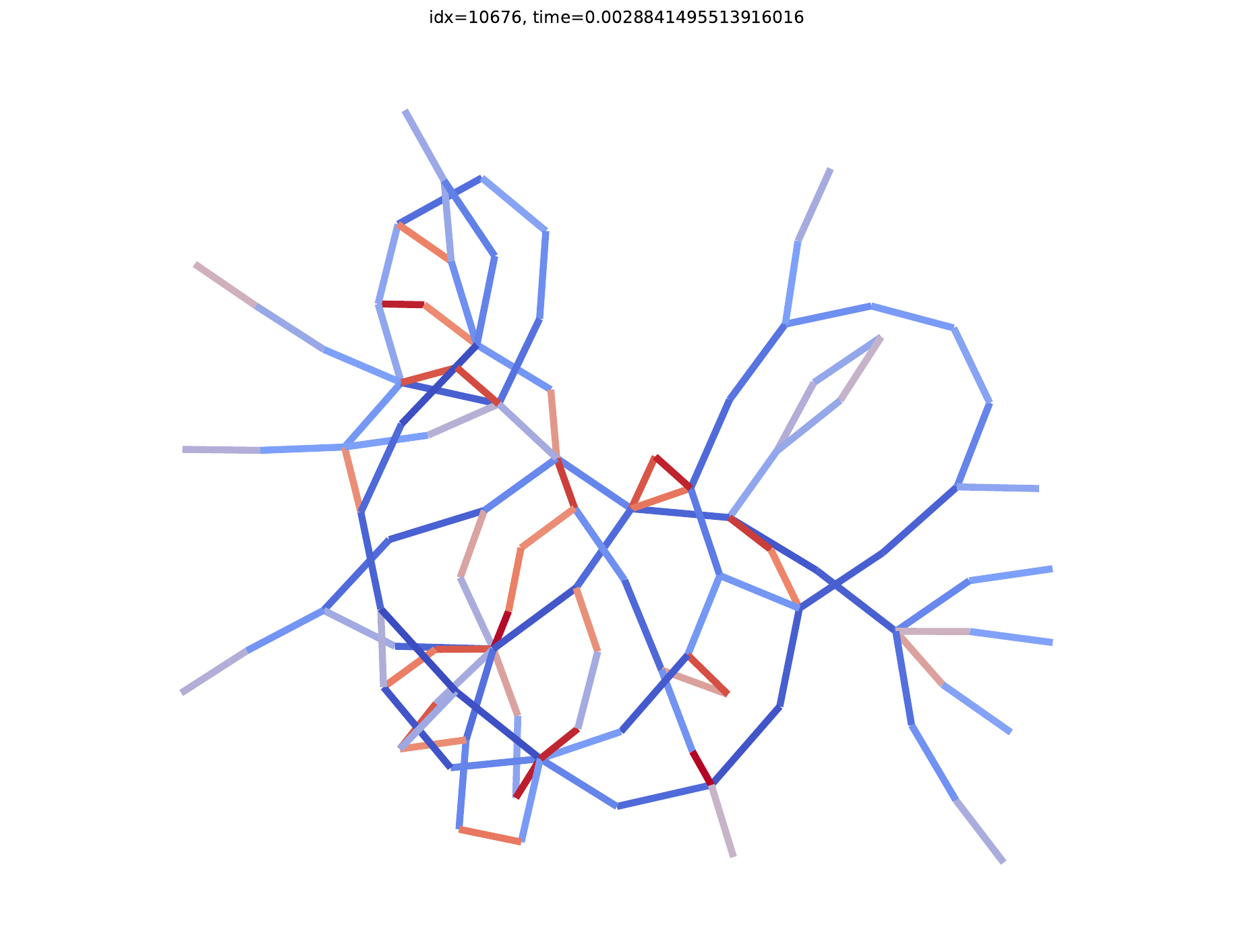} &
\imgcell{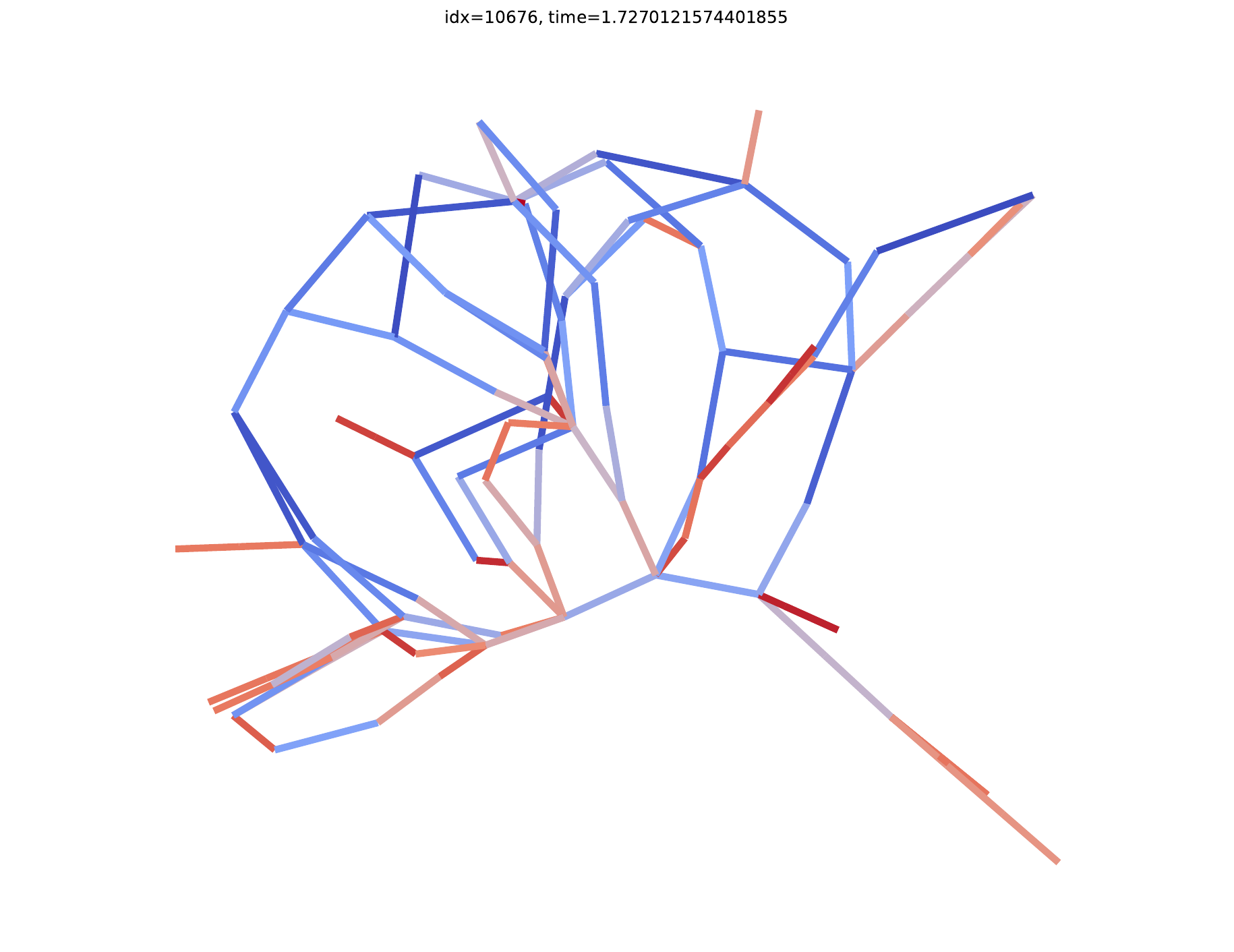} &
\imgcell{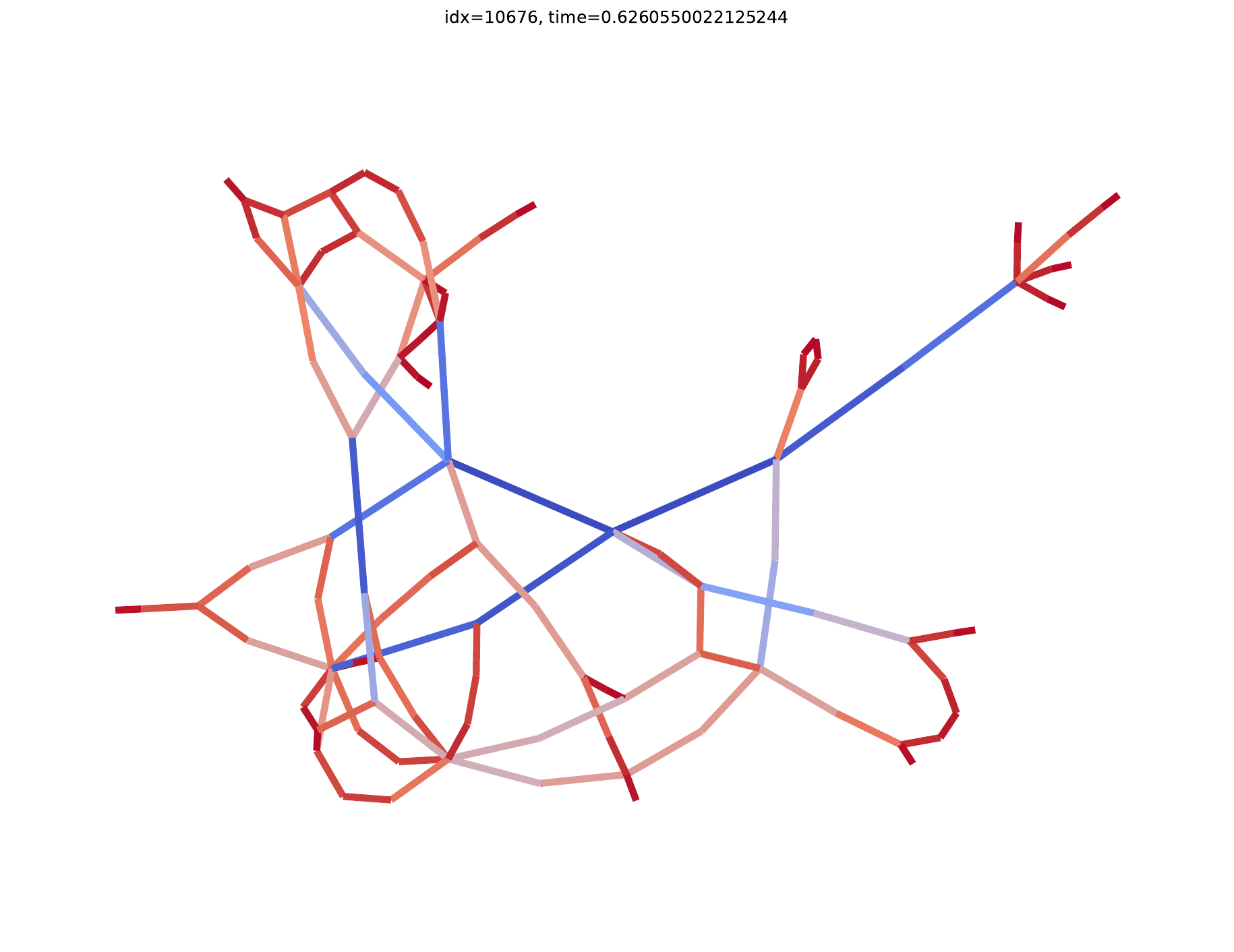} &
\imgcell{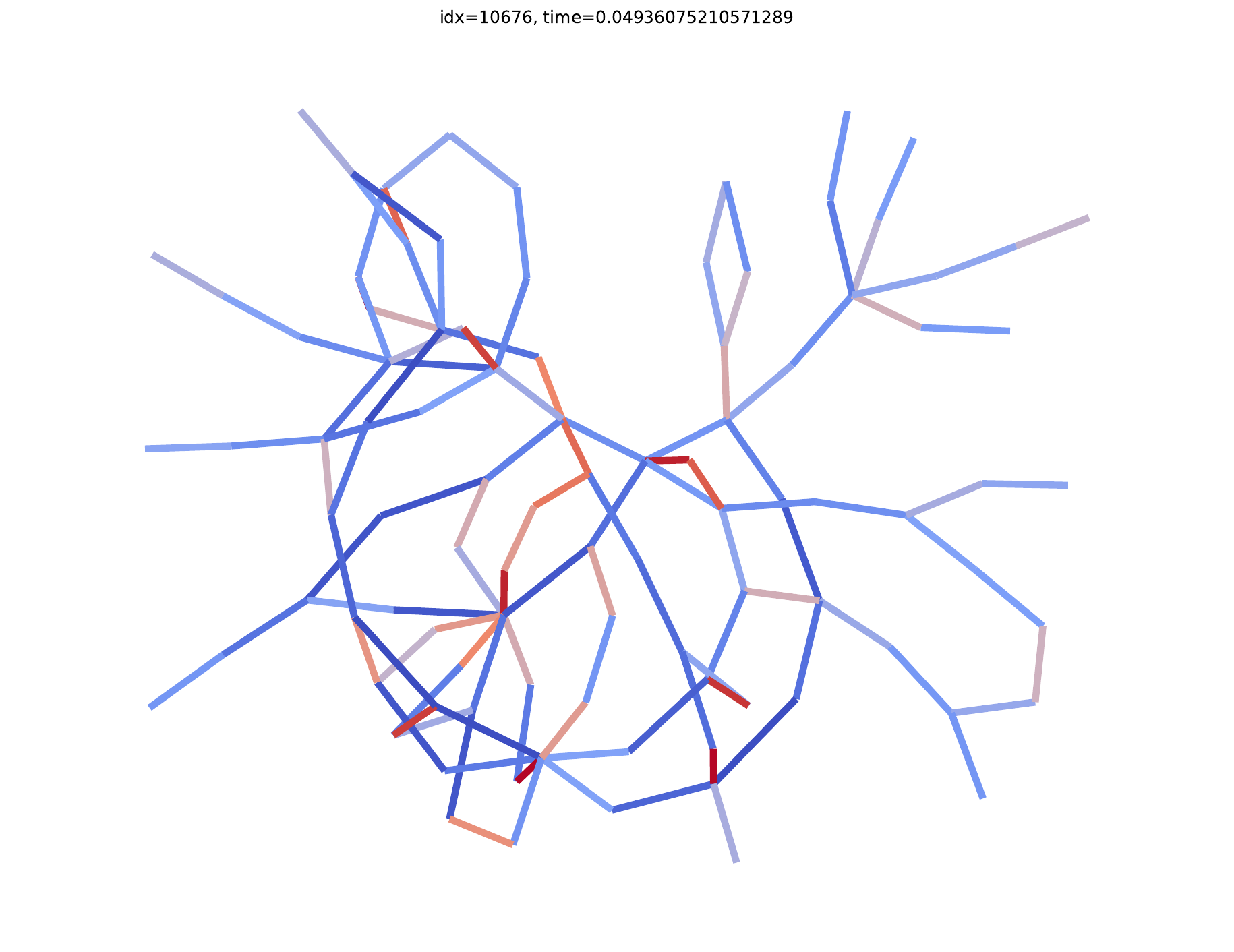} &
\imgcell{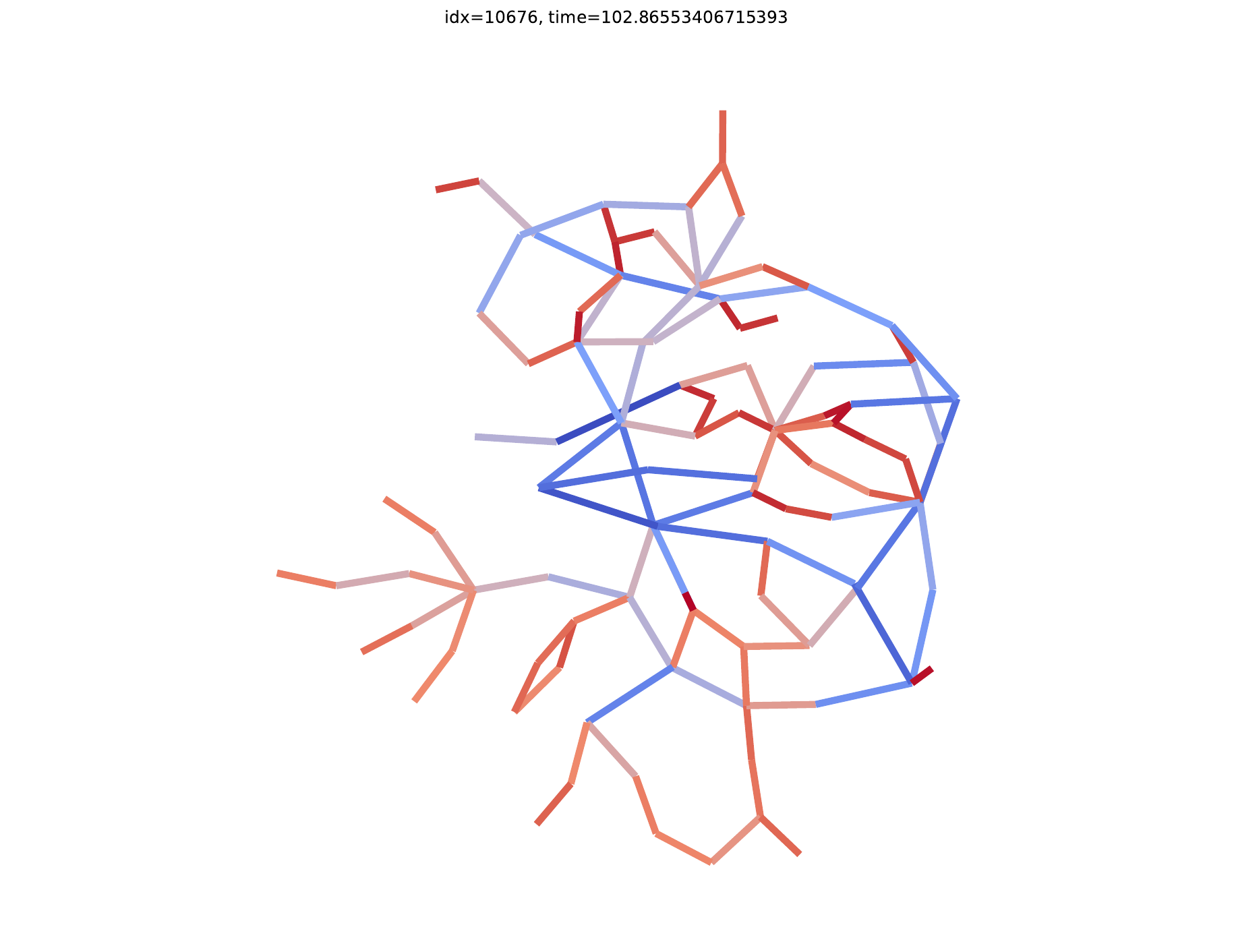} &
\imgcell{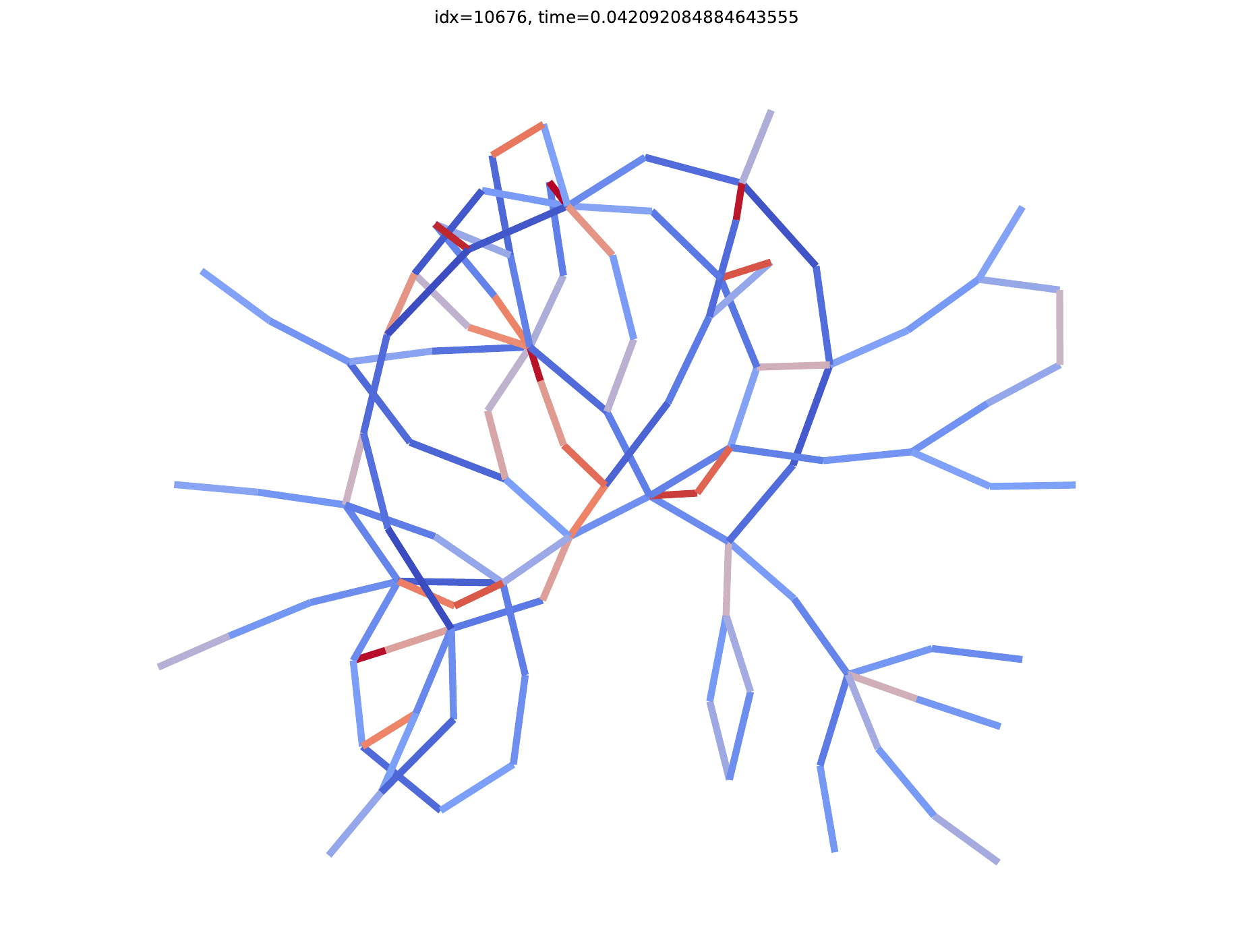} &
\imgcell{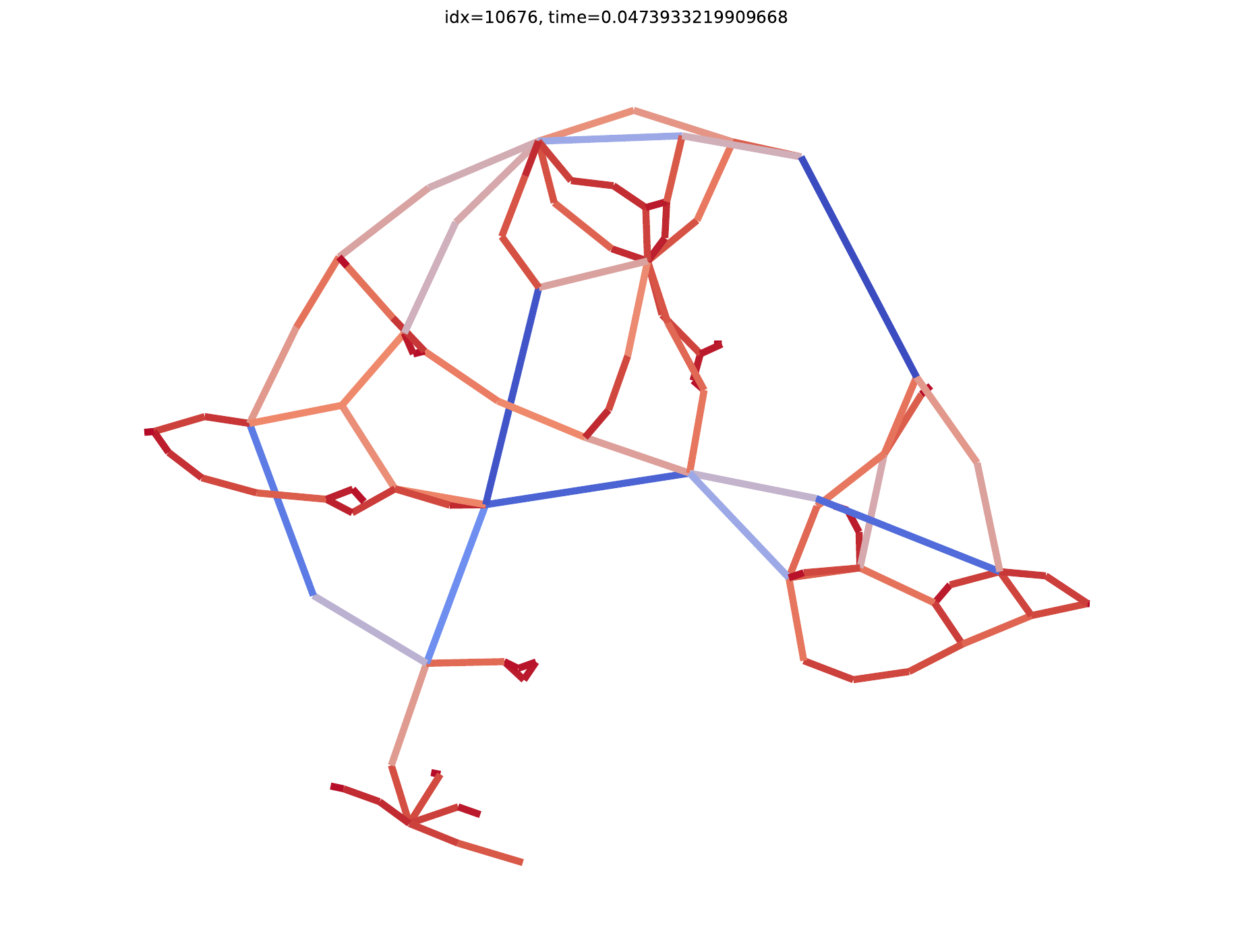} &
\imgcell{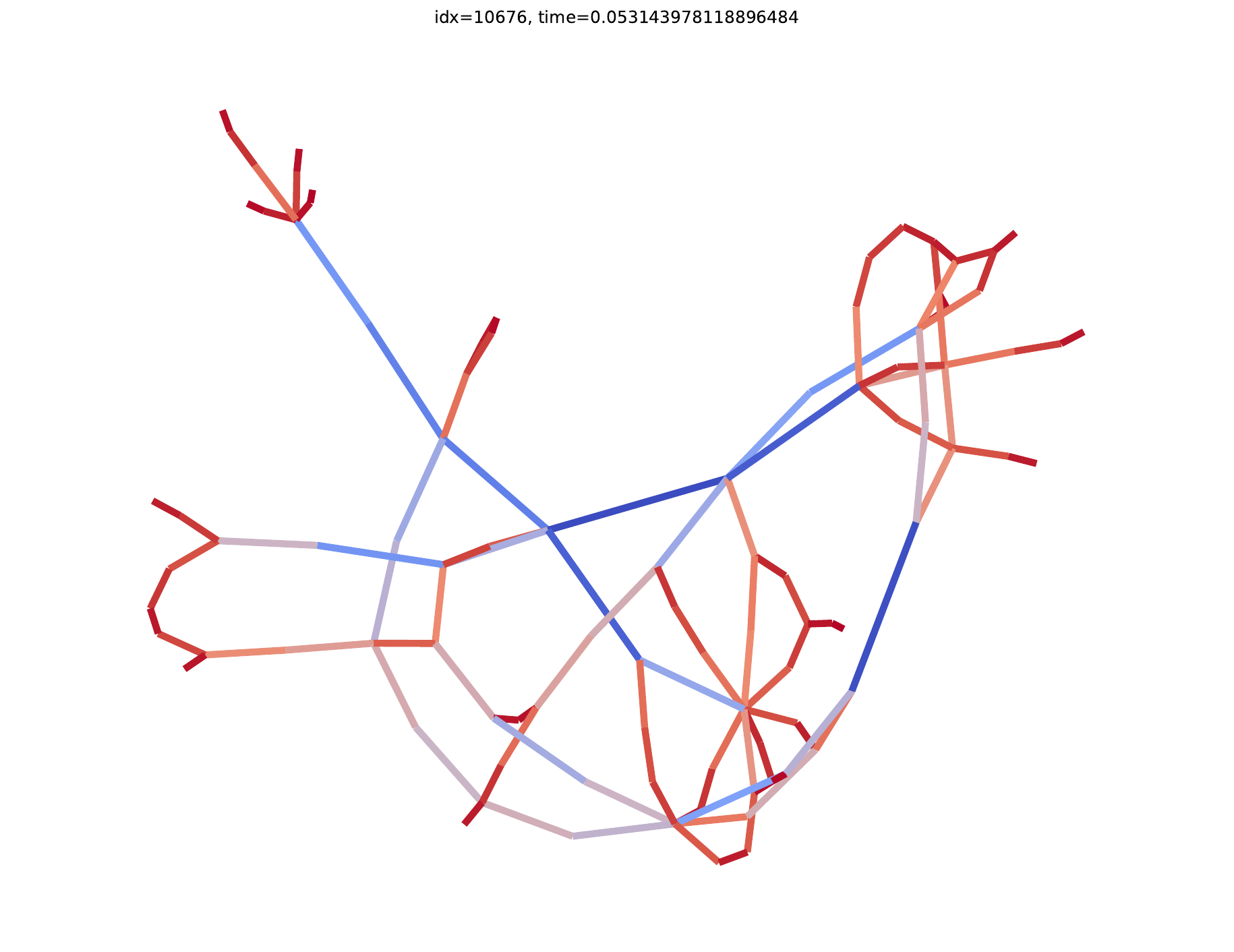} &
\imgcell{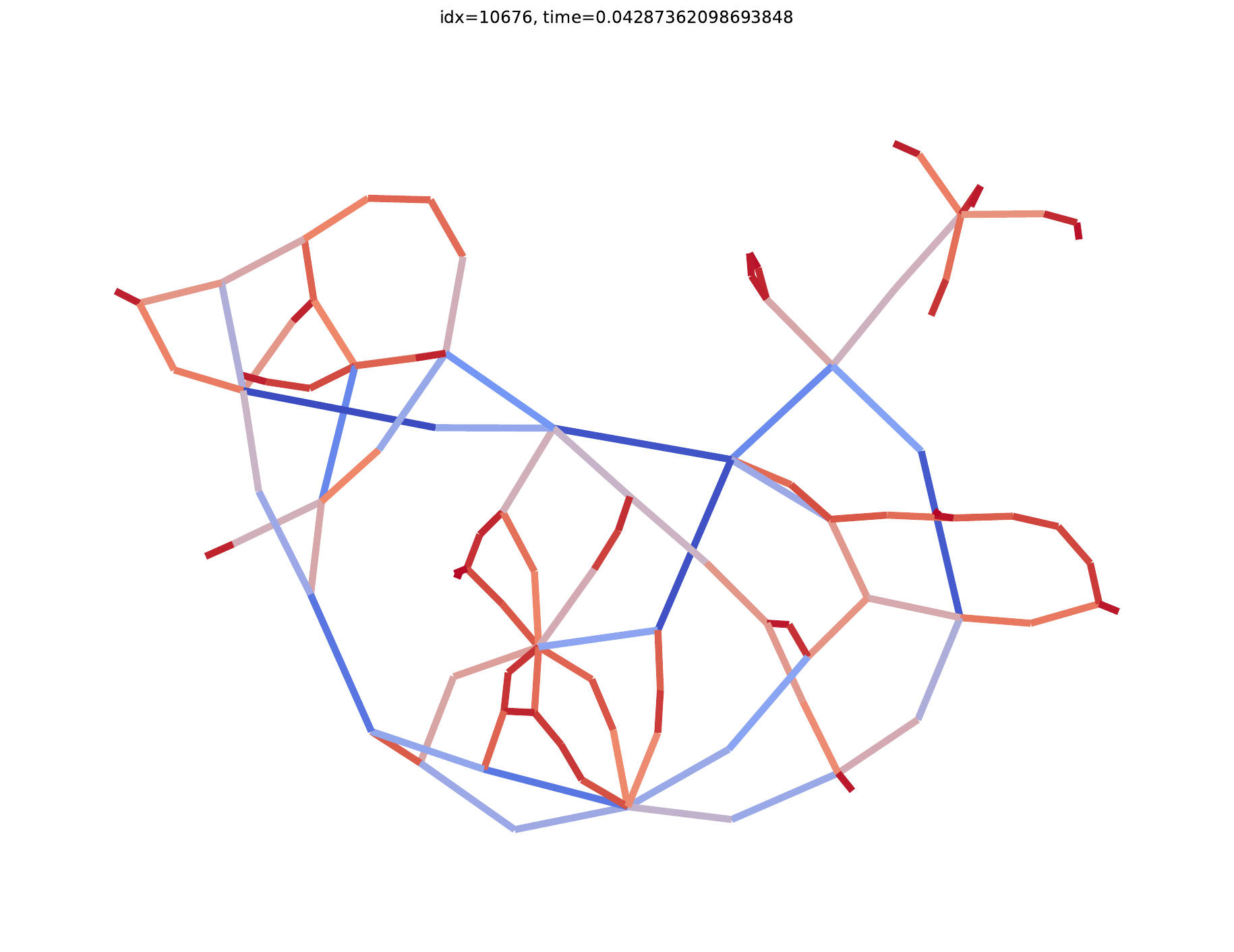} &
\imgcell{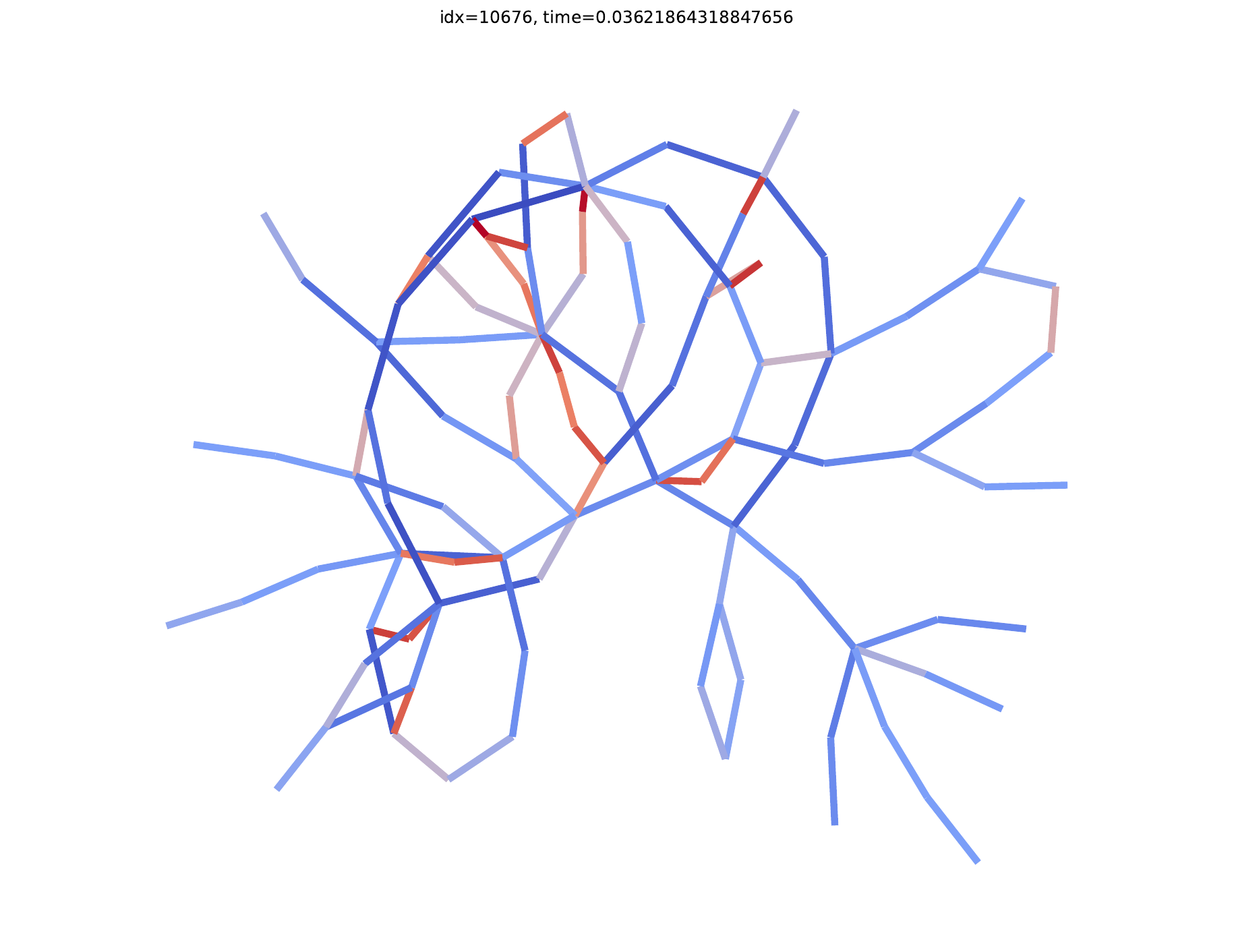} &
\imgcell{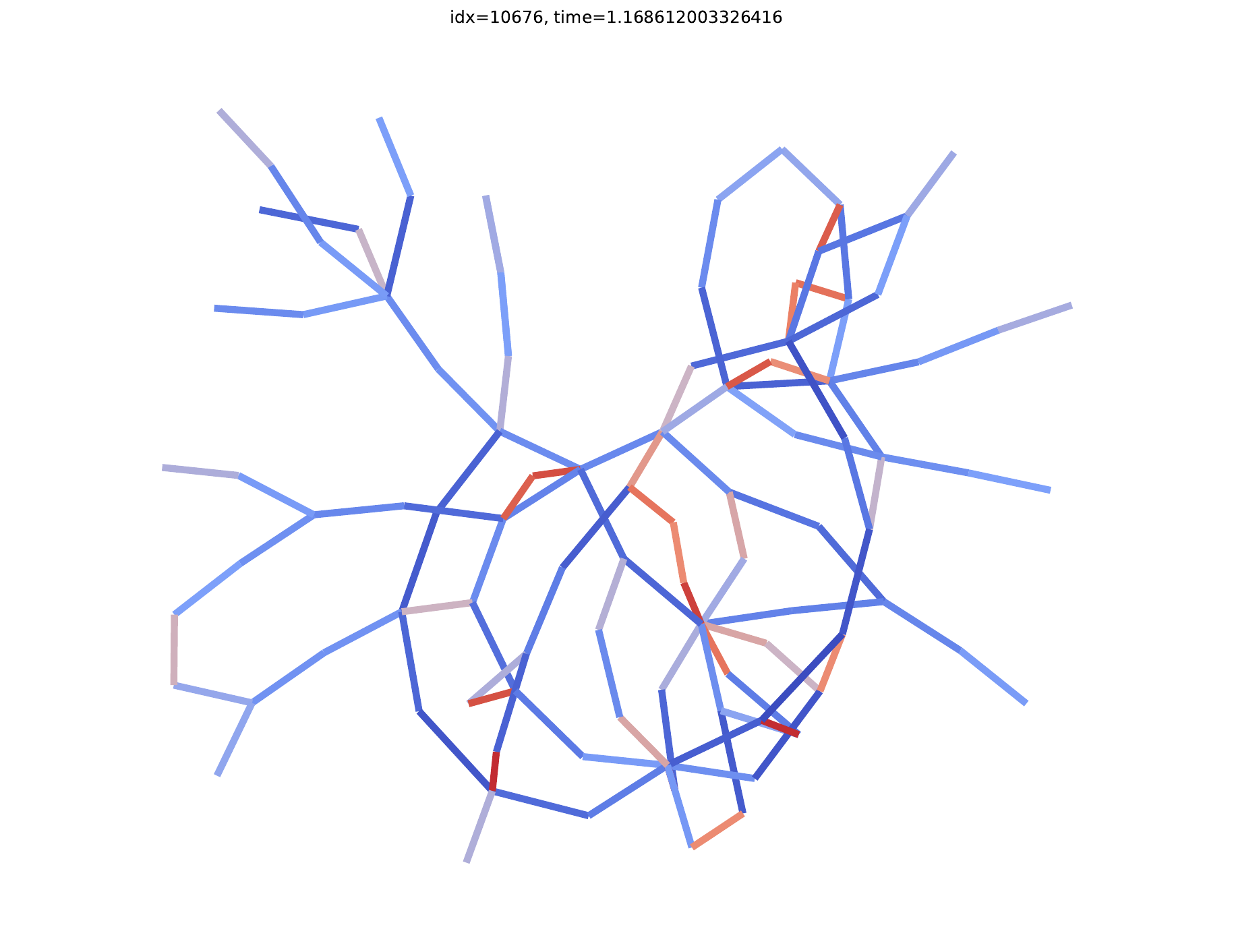} &
\imgcell{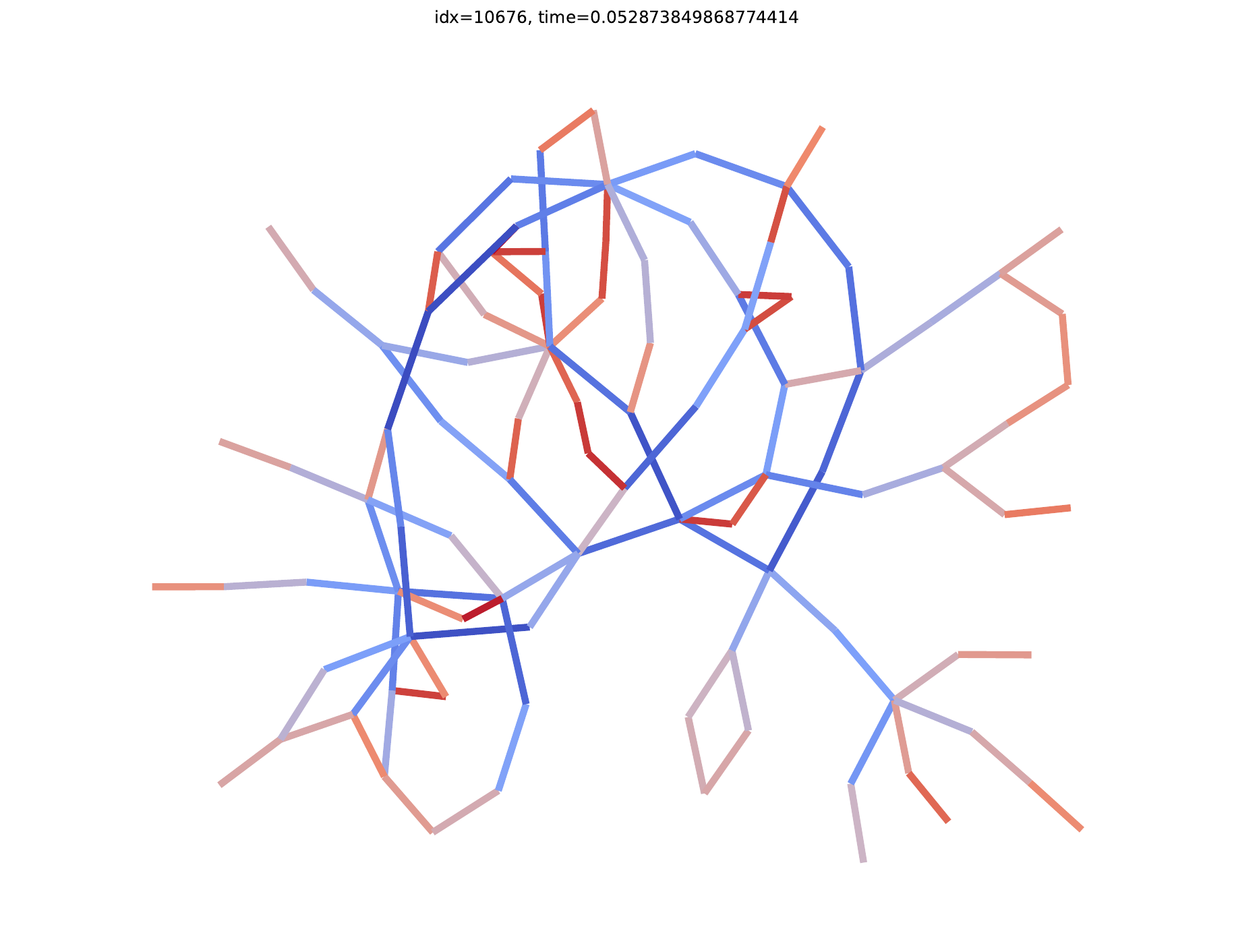} \\

&
t = 0.00s &
t = 1.73s &
t = 0.63s &
t = 0.05s &
t = 102.87s &
t = 0.04s &
t = 0.05s &
t = 0.05s &
t = 0.04s &
t = 0.04s &
t = 0.06s &
t = 0.05s \\

\makecell{\bfseries grafo9582.80\\N = 63\\M = 88} &
\imgcell{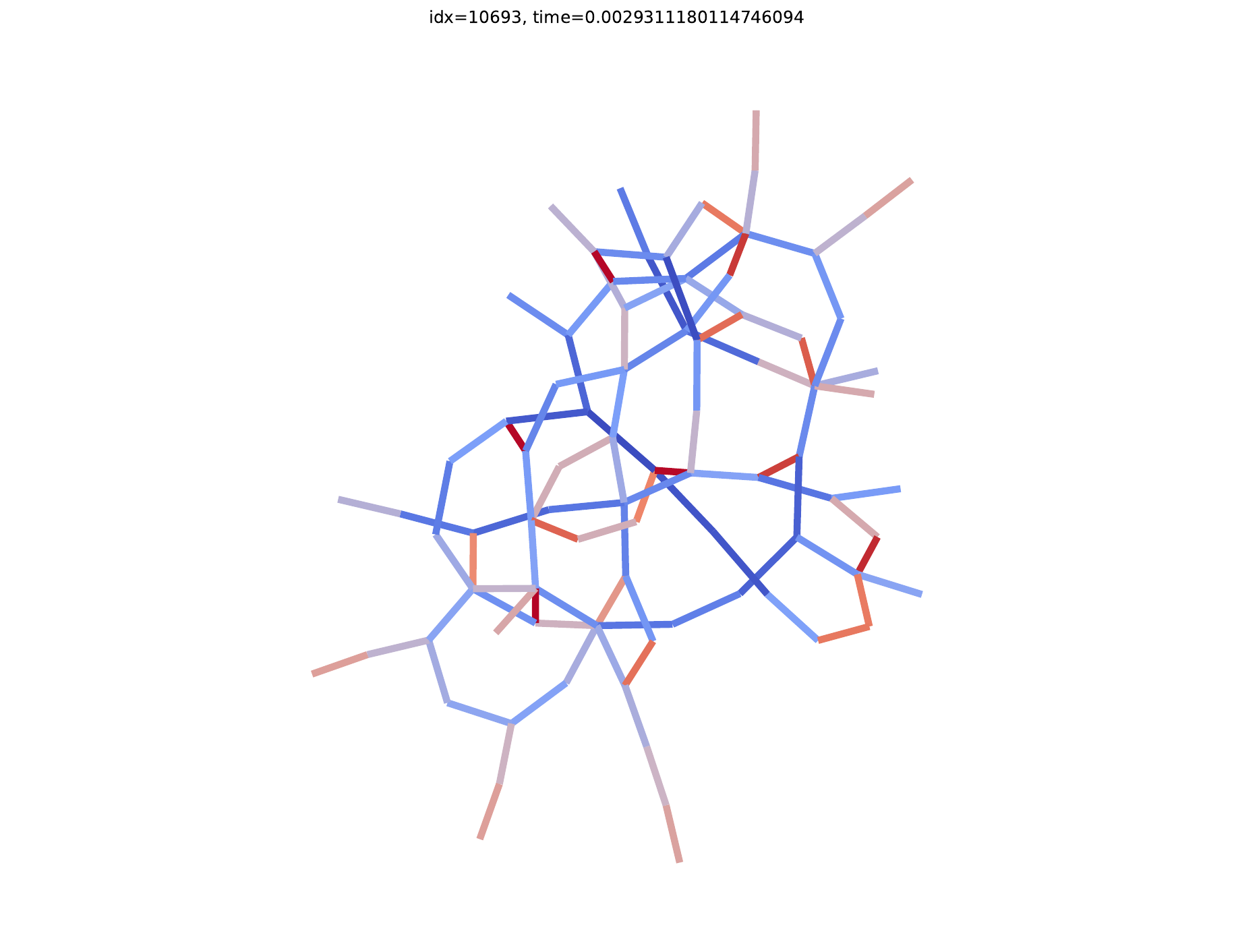} &
\imgcell{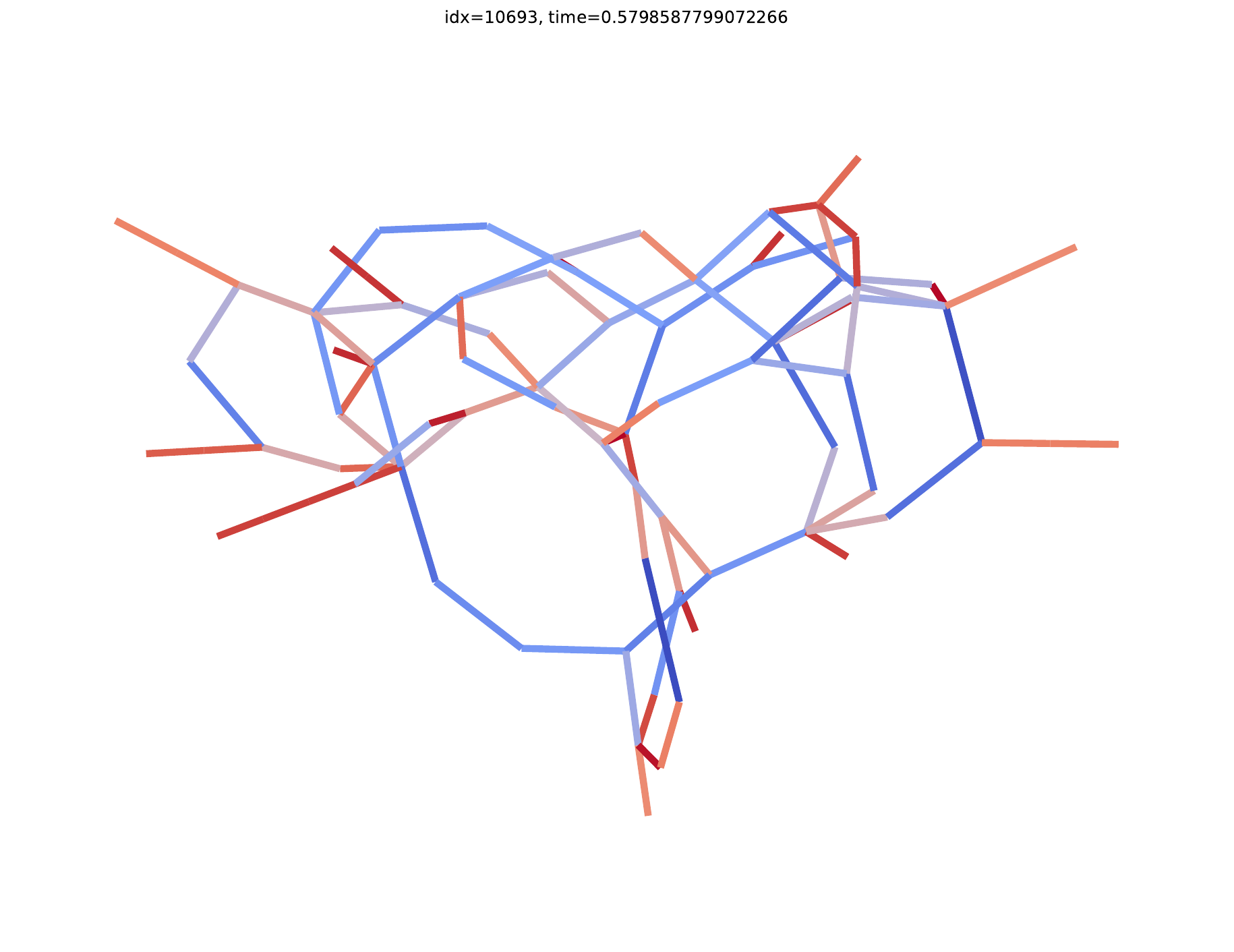} &
\imgcell{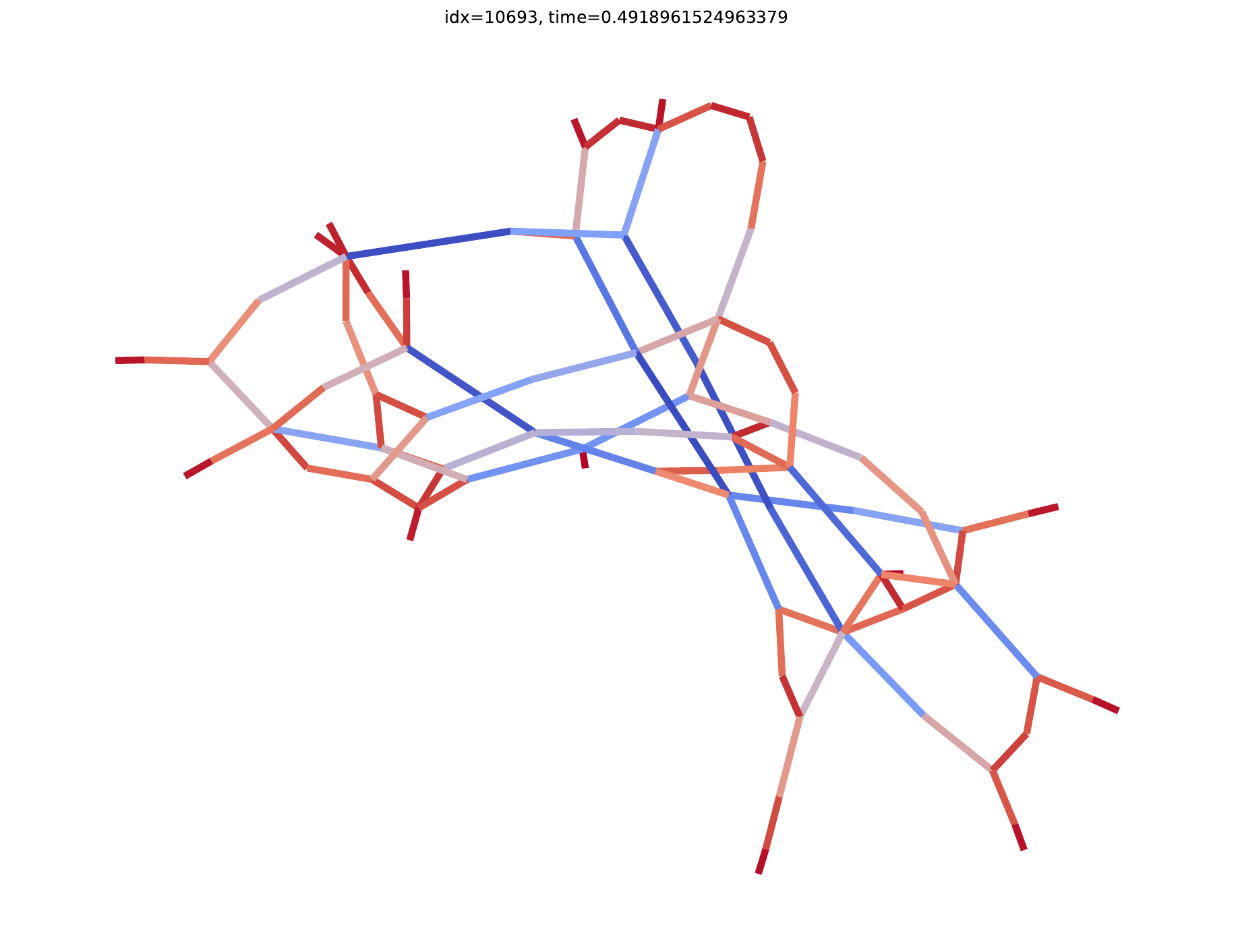} &
\imgcell{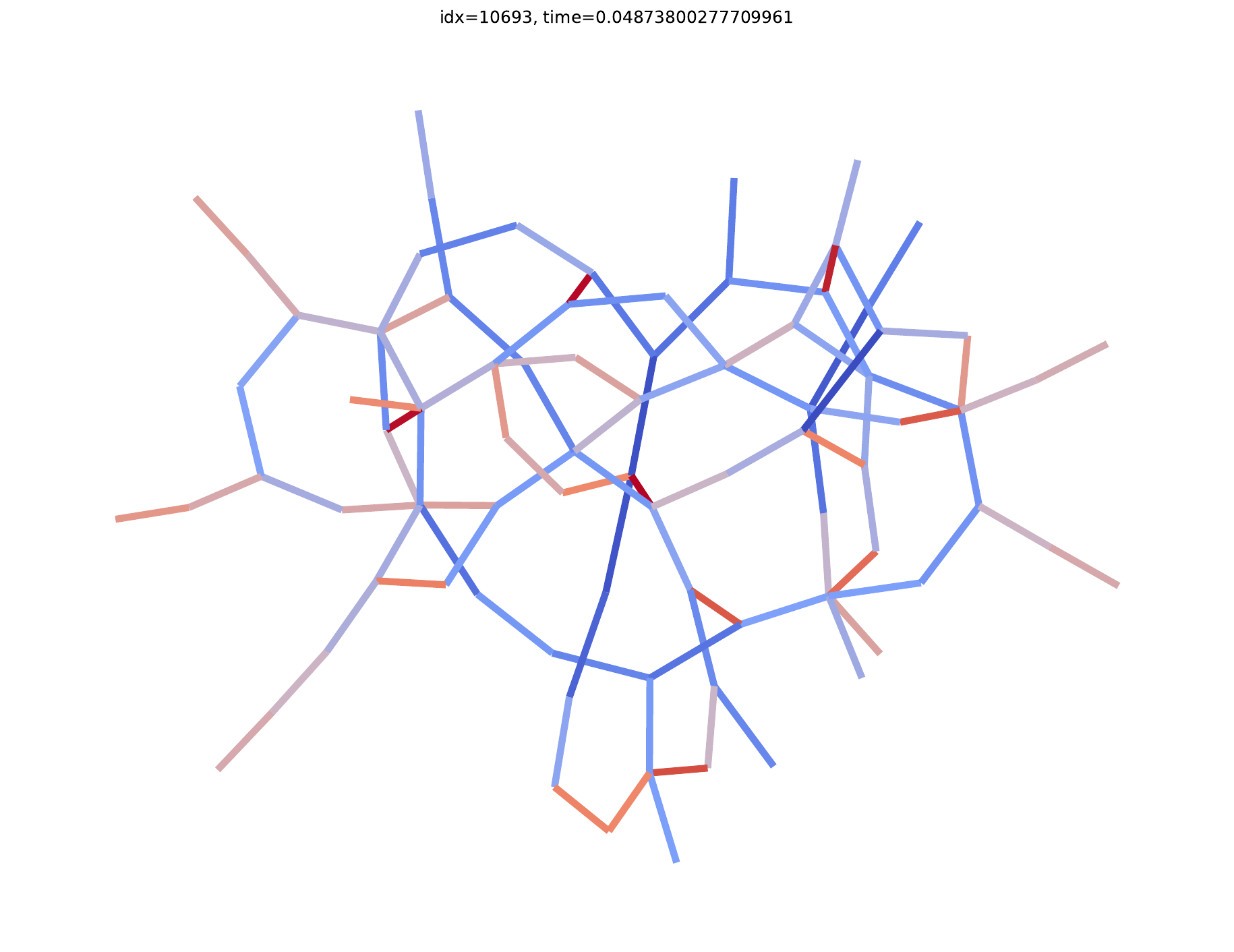} &
\imgcell{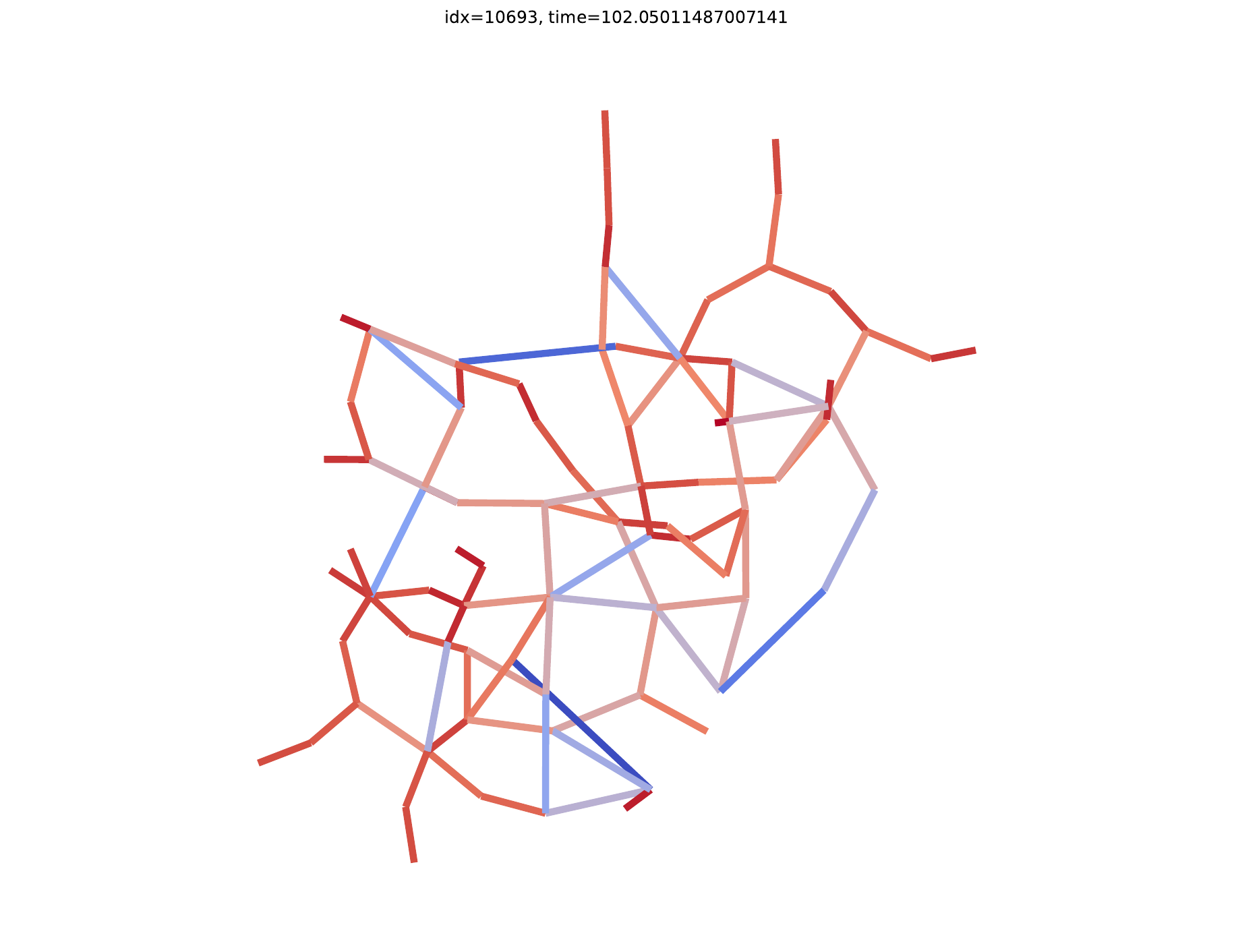} &
\imgcell{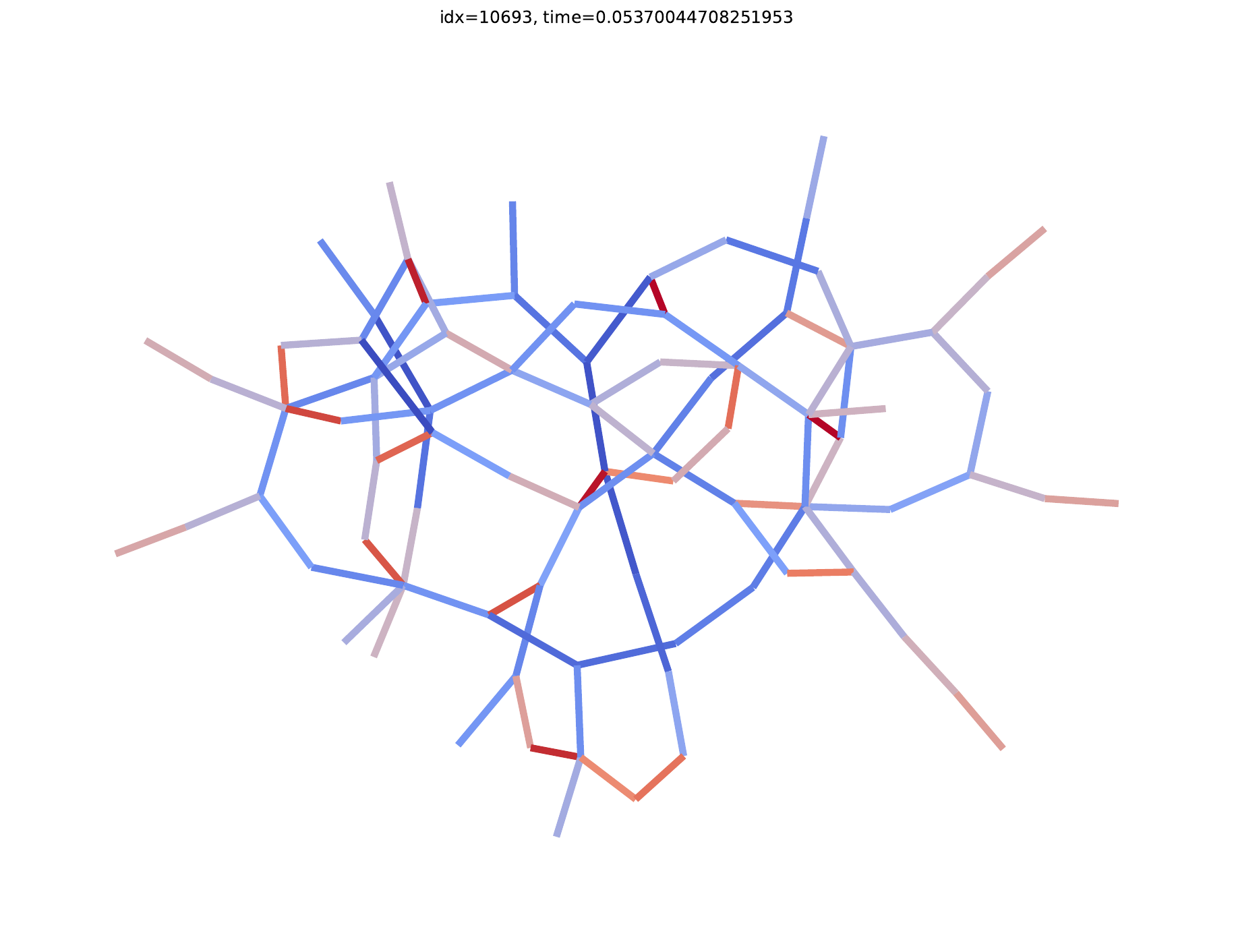} &
\imgcell{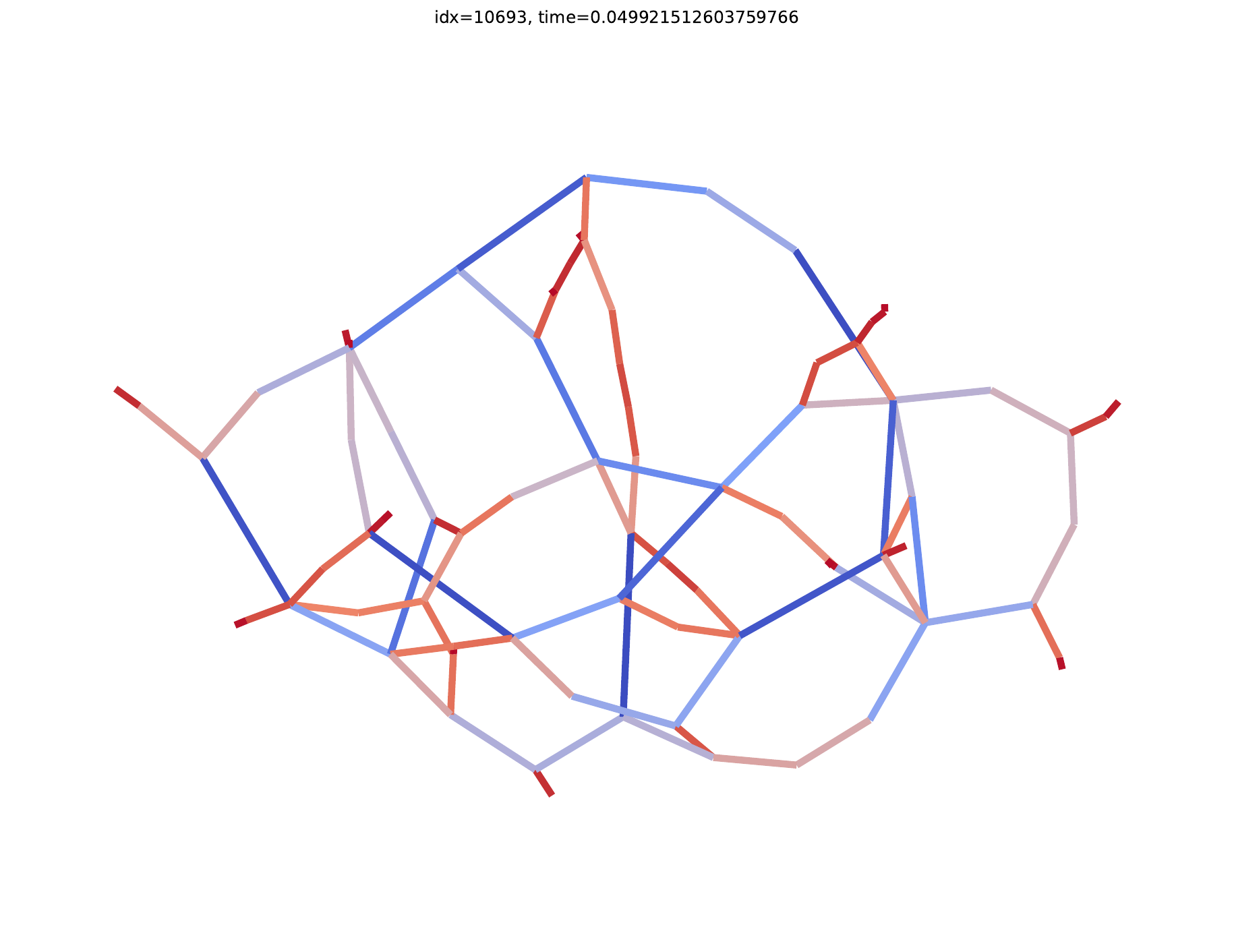} &
\imgcell{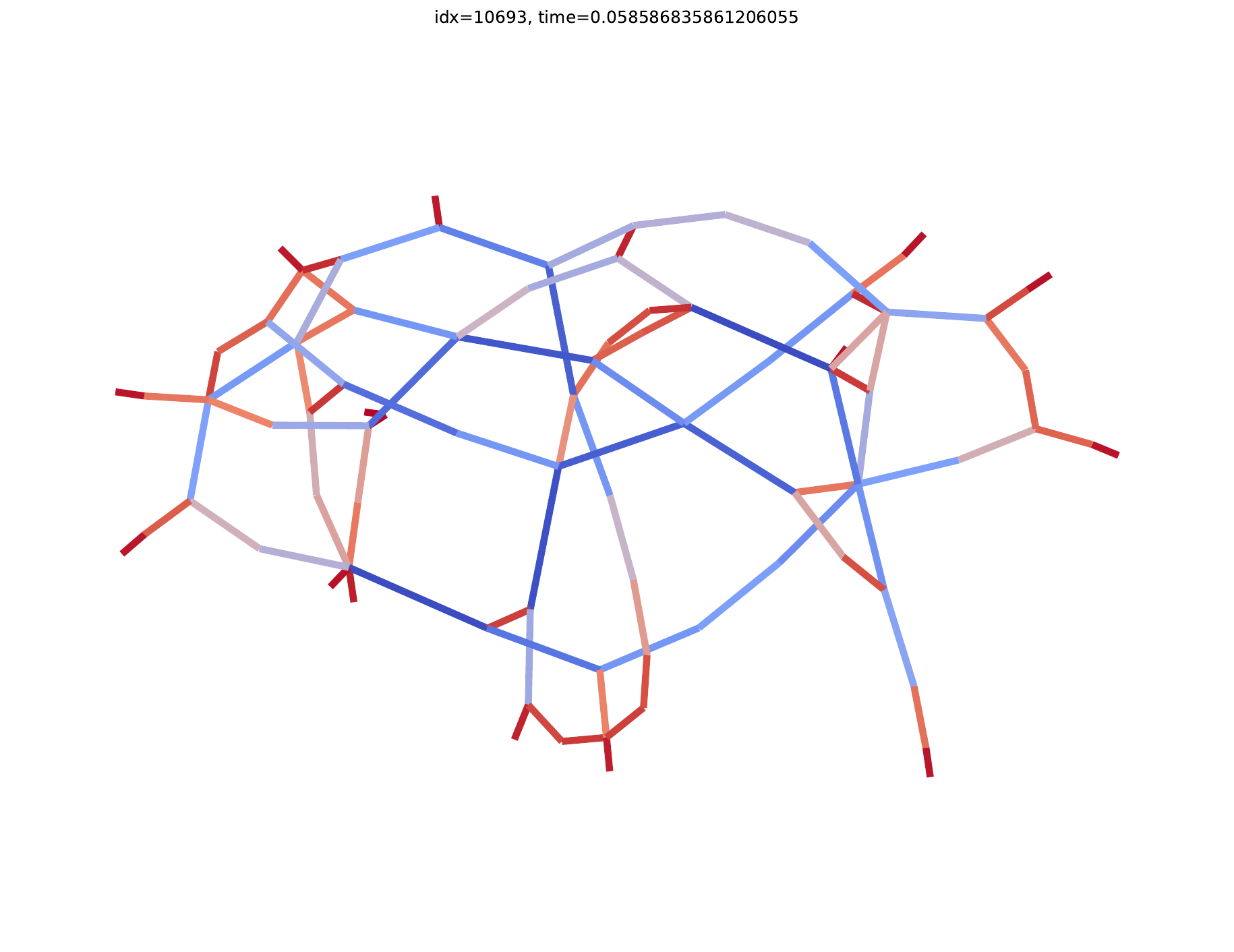} &
\imgcell{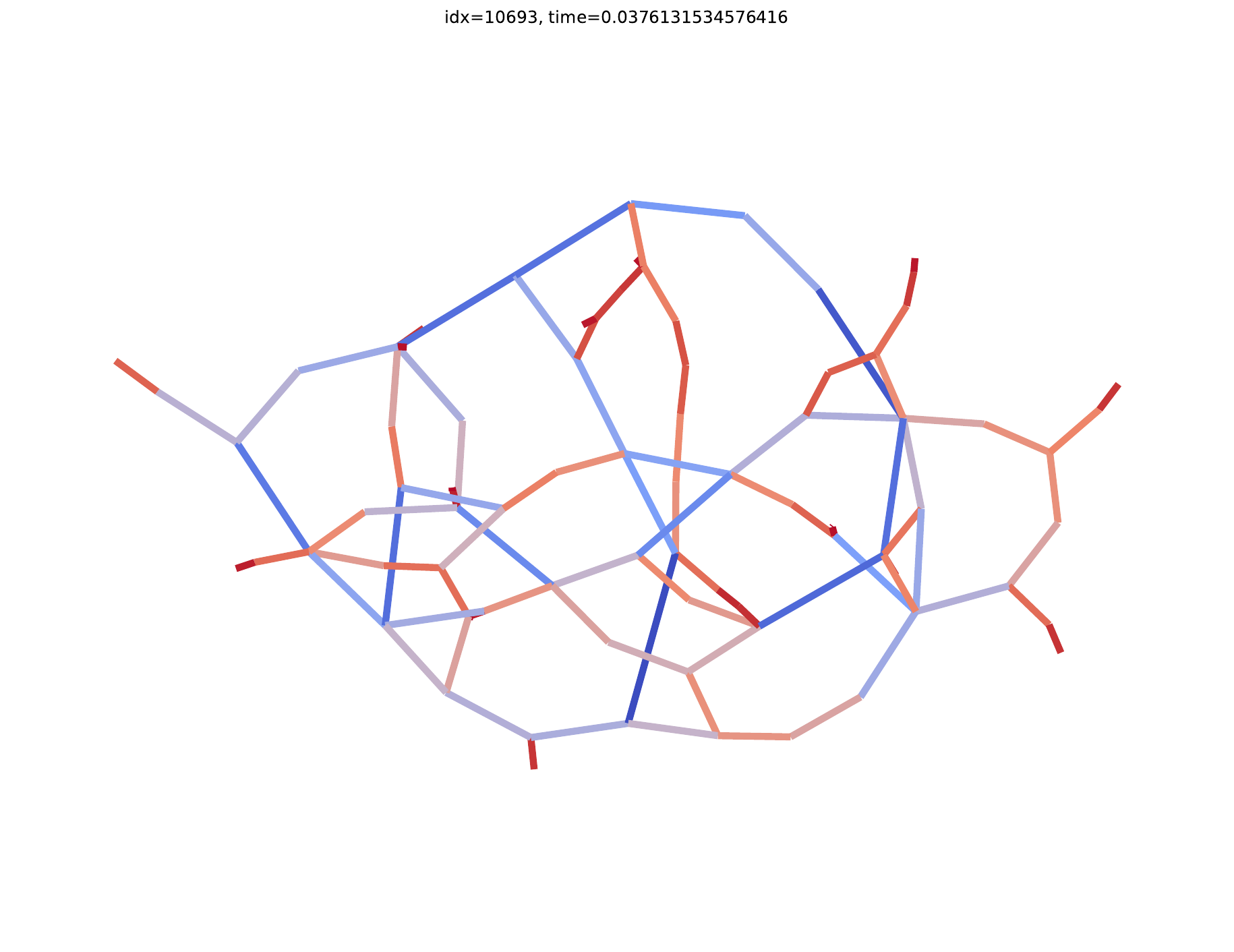} &
\imgcell{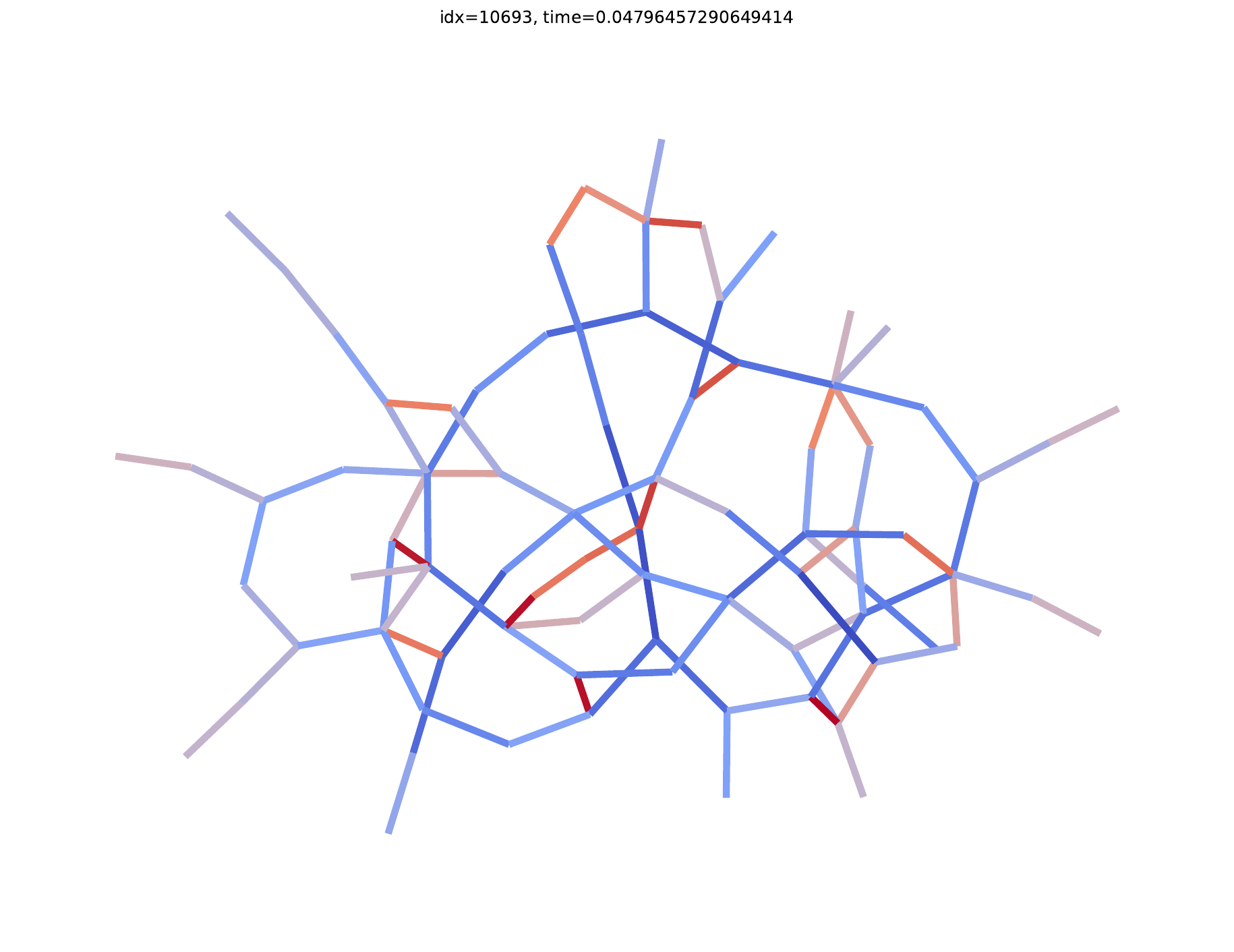} &
\imgcell{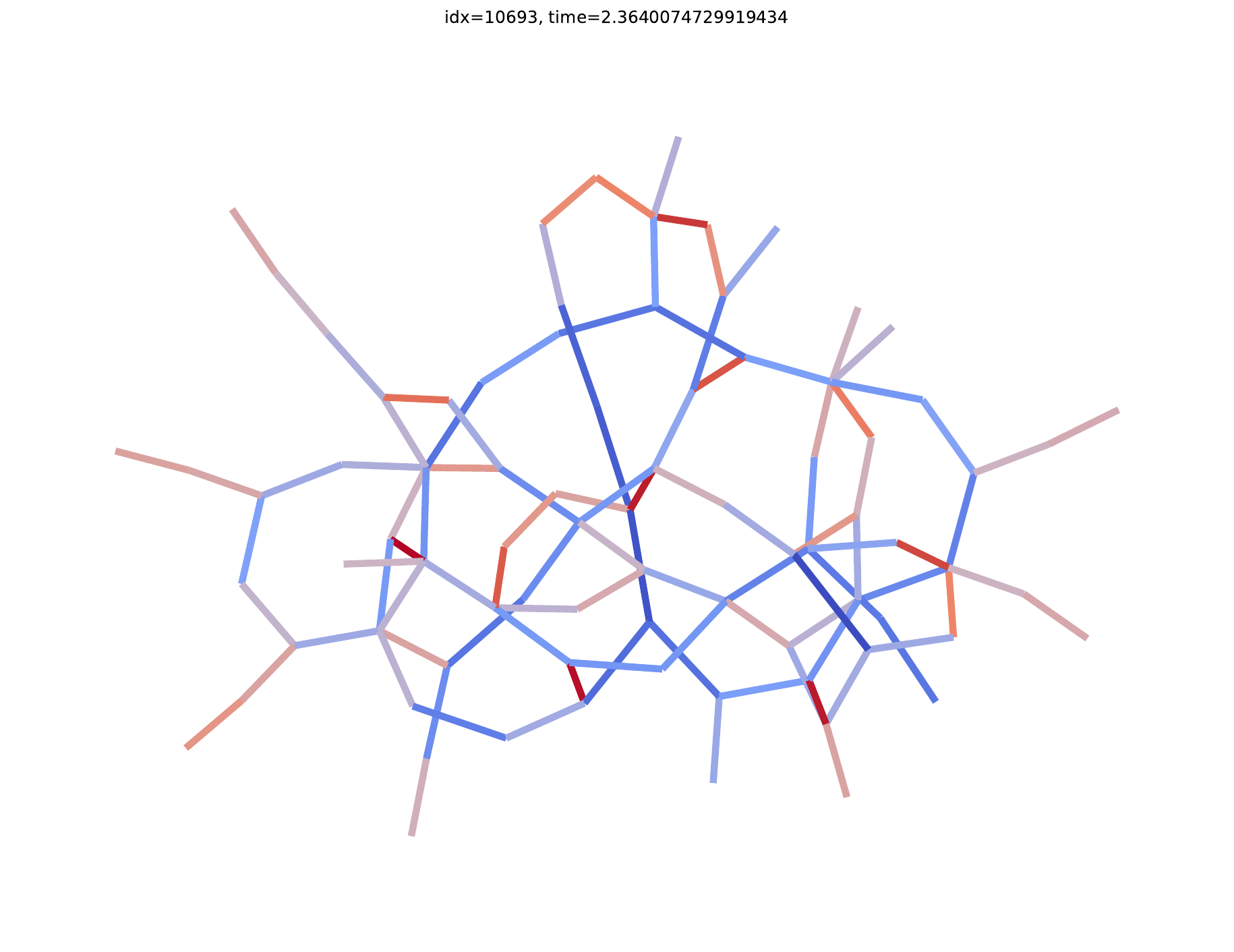} &
\imgcell{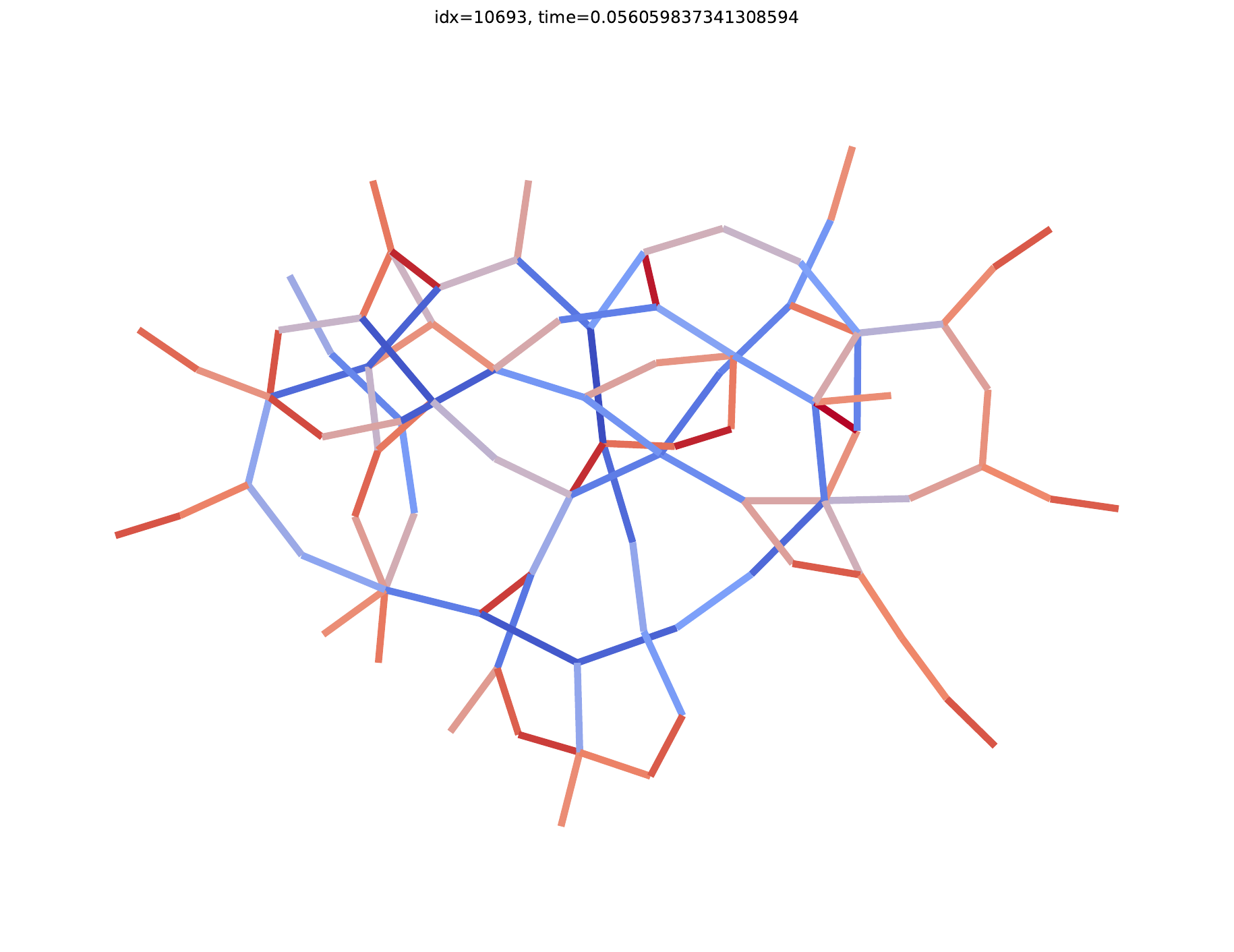} \\

&
t = 0.00s &
t = 0.58s &
t = 0.49s &
t = 0.05s &
t = 102.05s &
t = 0.05s &
t = 0.05s &
t = 0.06s &
t = 0.04s &
t = 0.05s &
t = 0.06s &
t = 0.06s \\

\makecell{\bfseries grafo1005.11\\N = 16\\M = 20} &
\imgcell{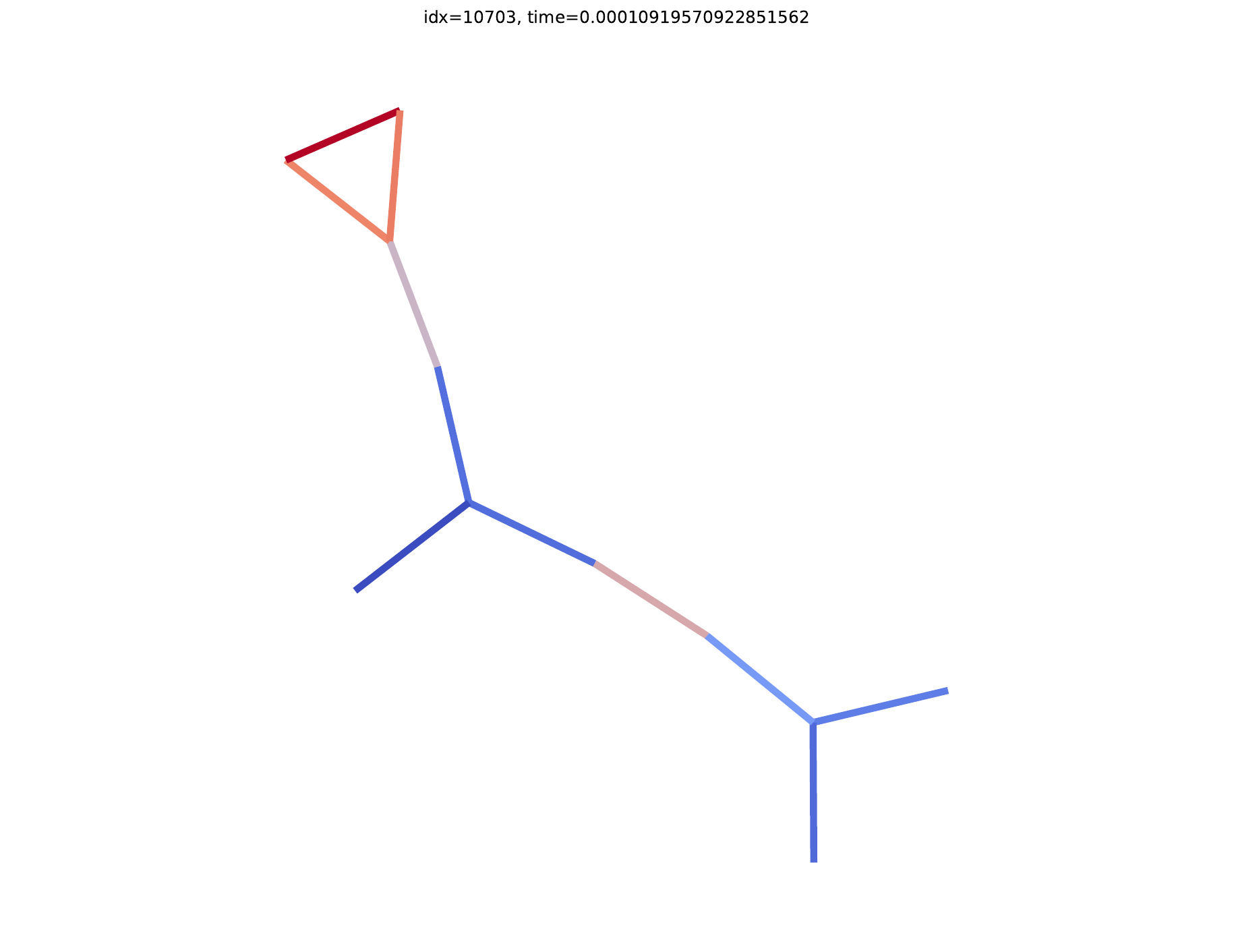} &
\imgcell{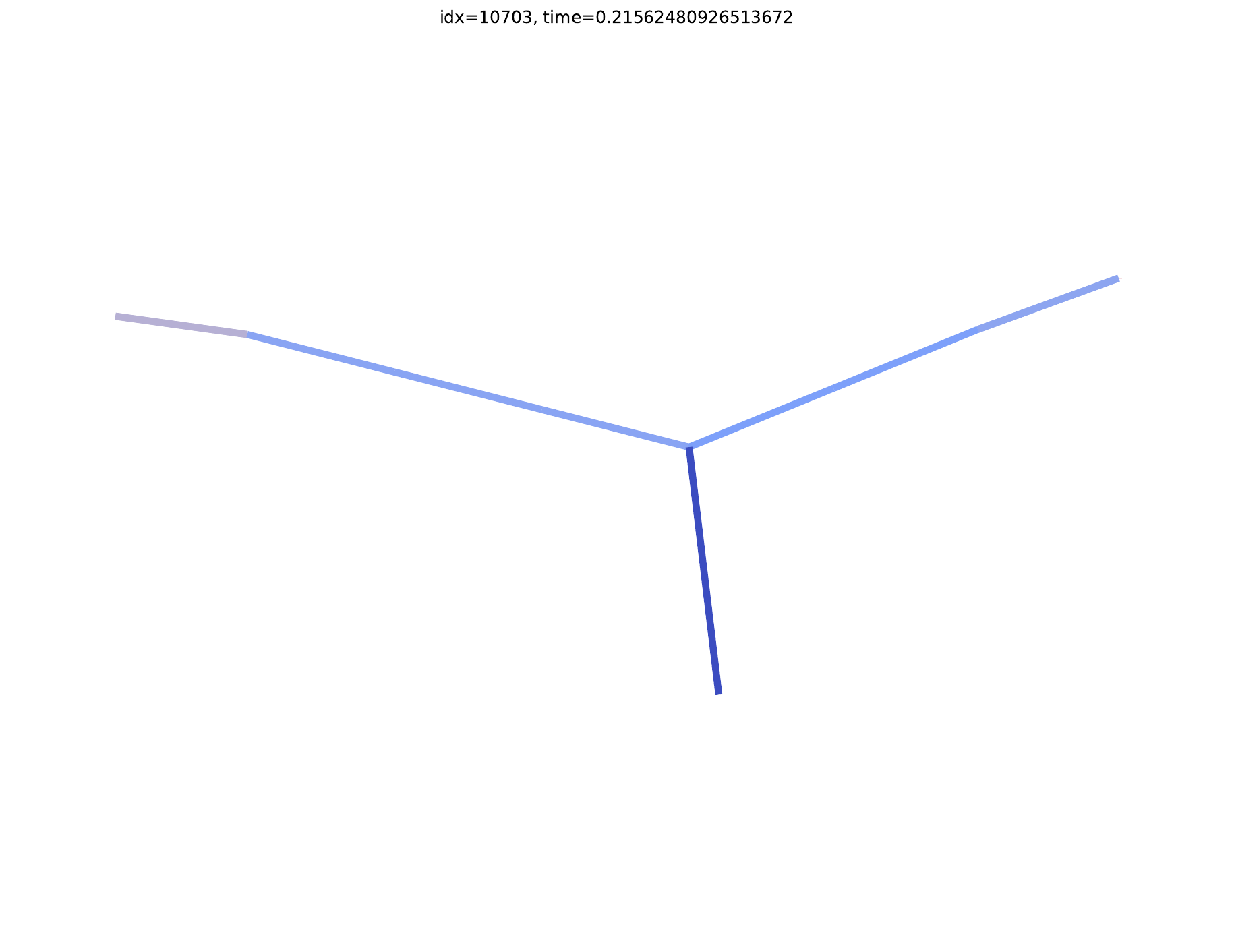} &
\imgcell{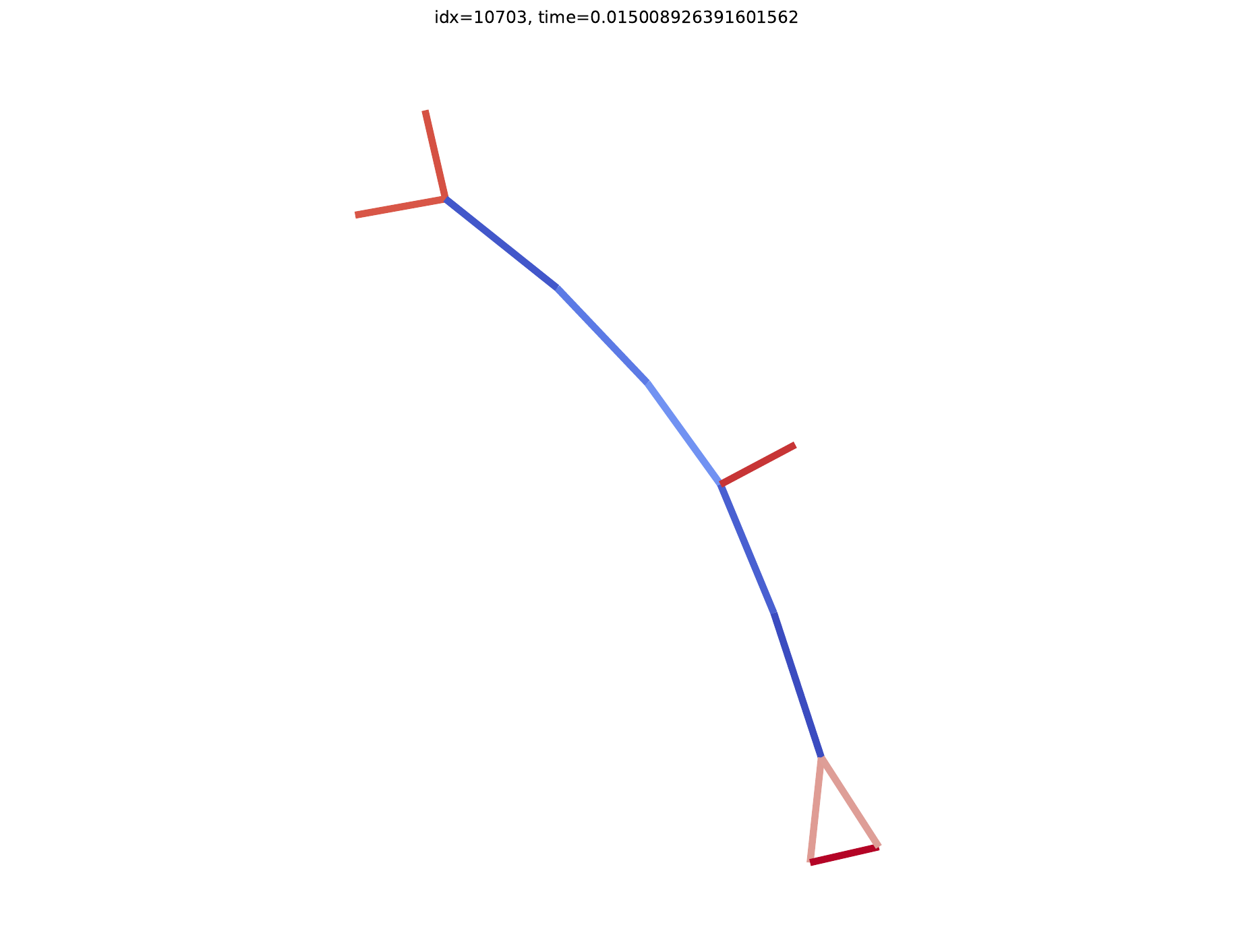} &
\imgcell{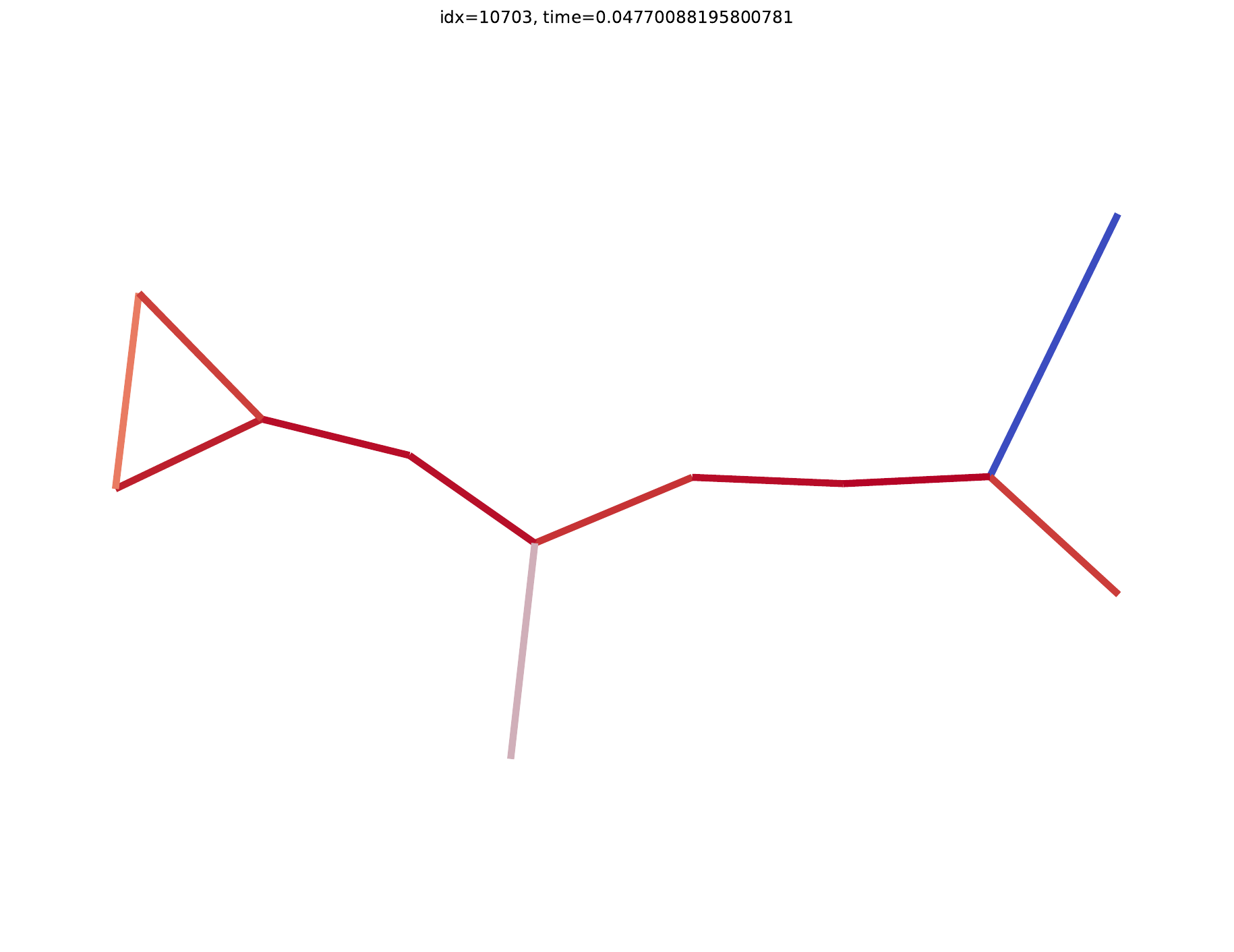} &
\imgcell{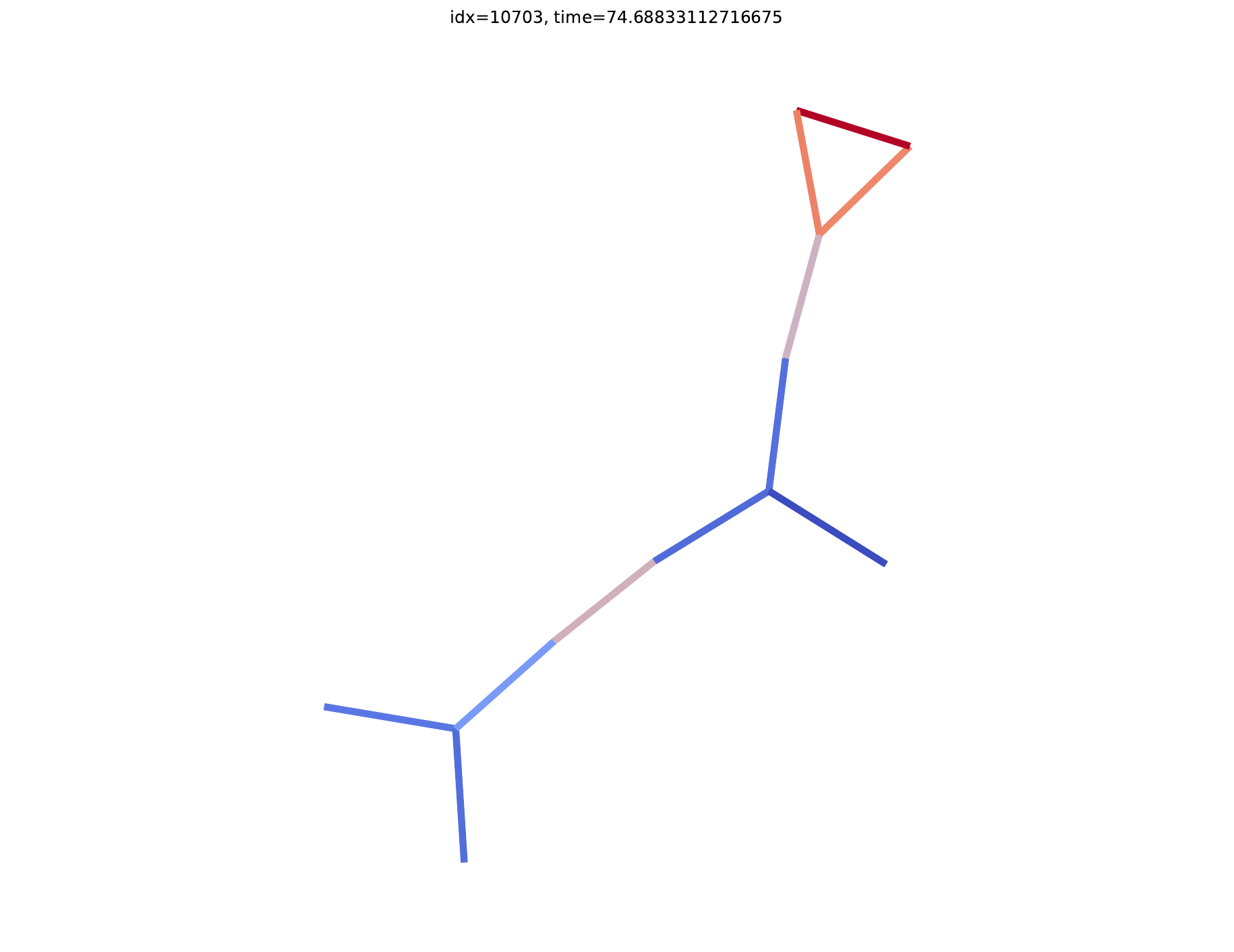} &
\imgcell{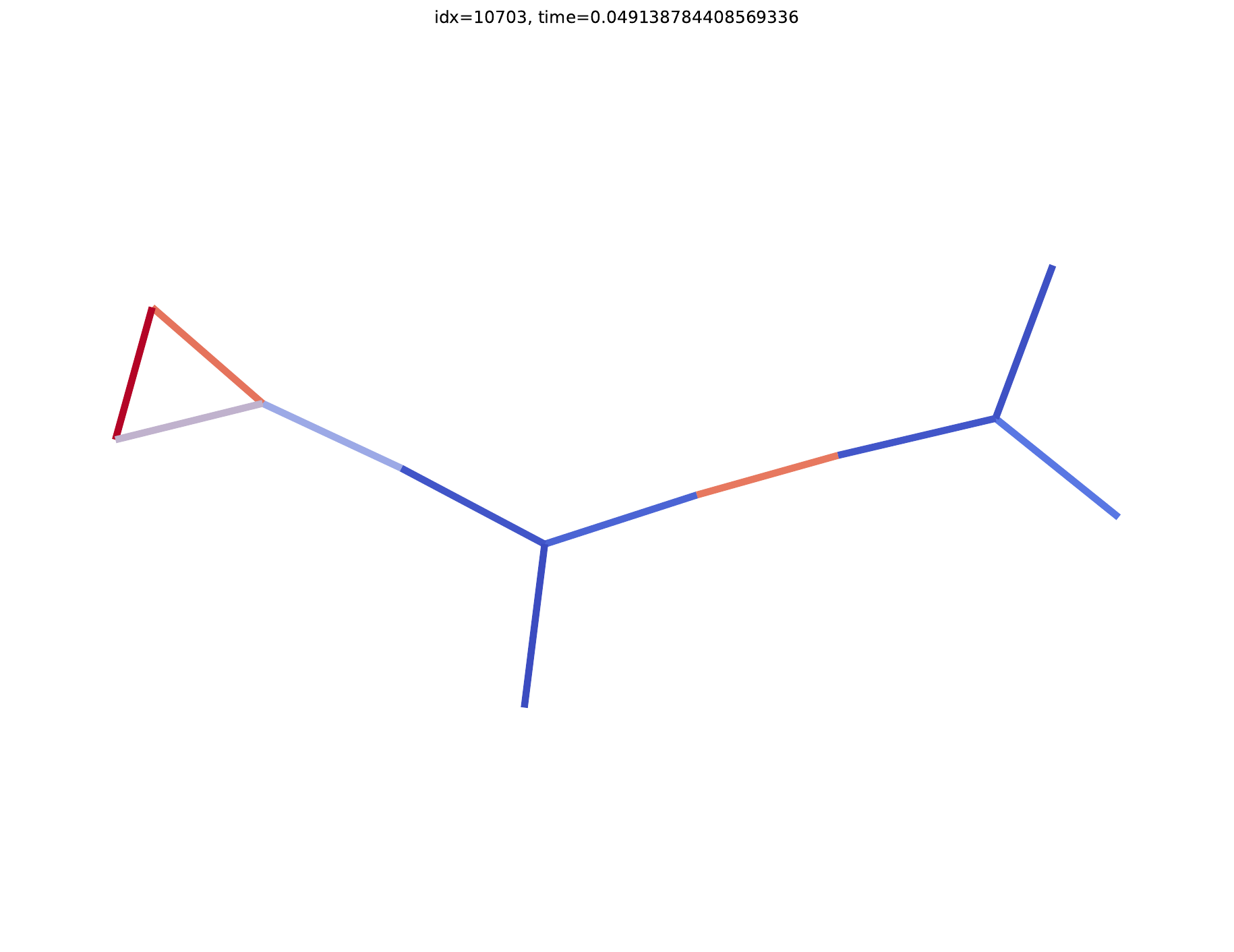} &
\imgcell{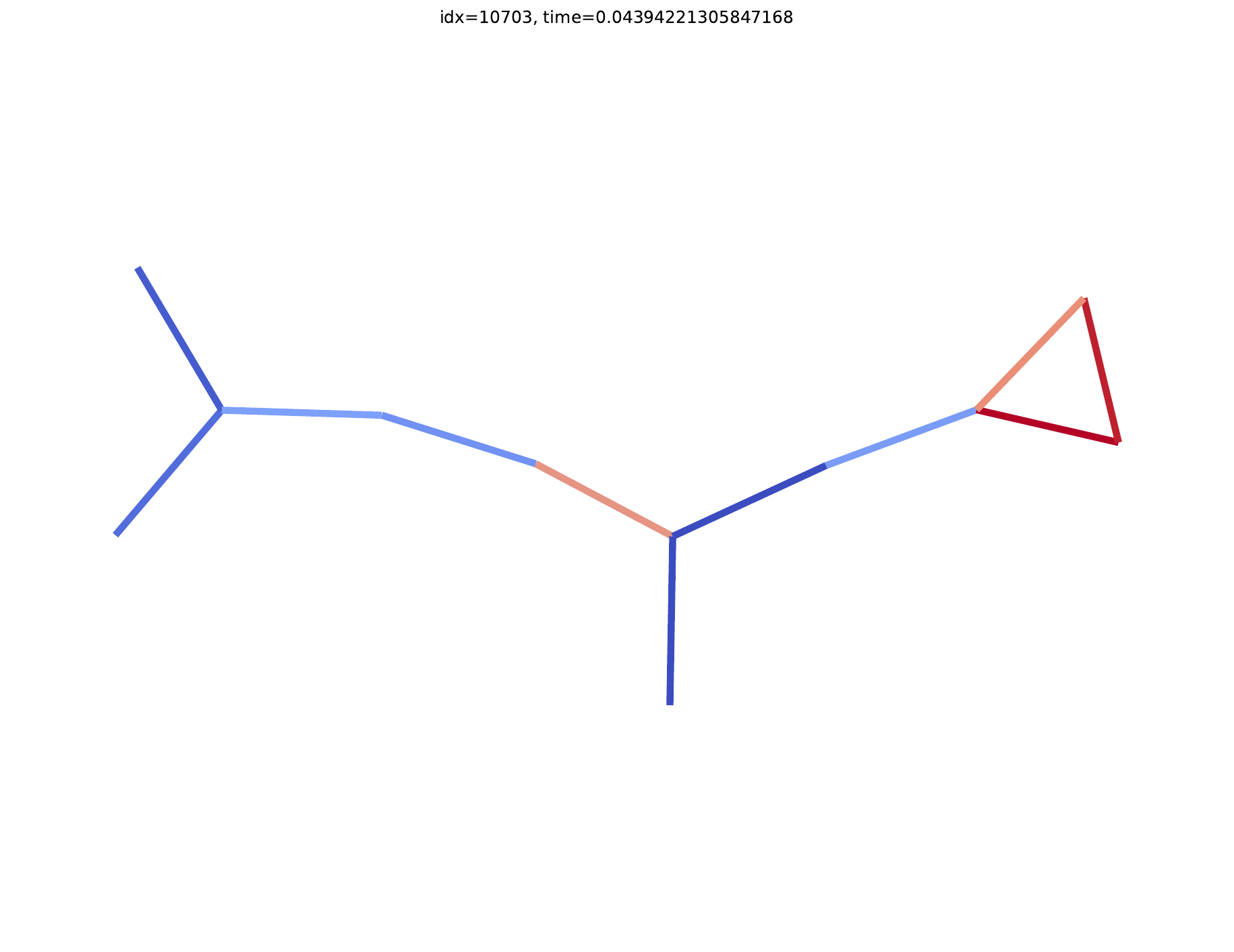} &
\imgcell{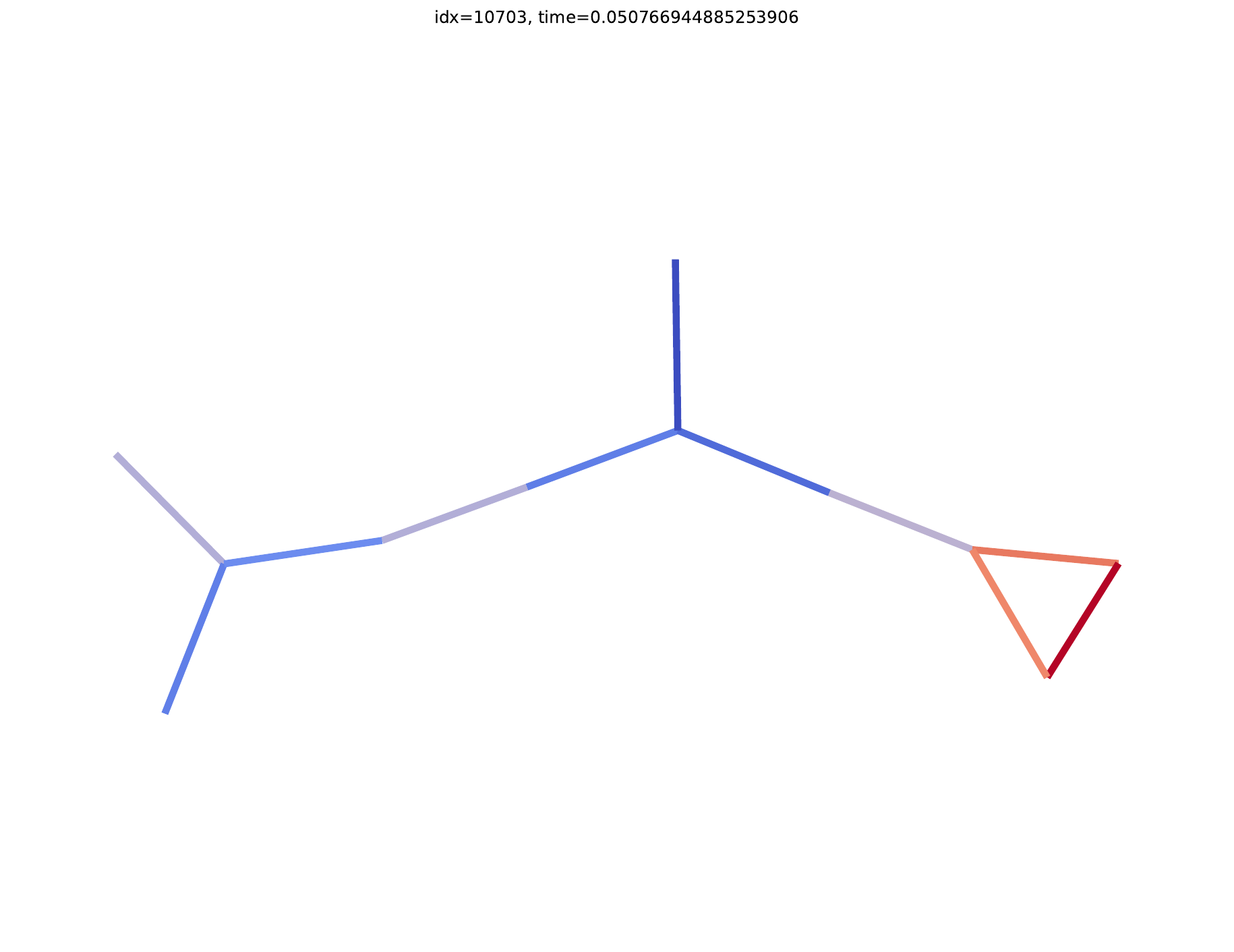} &
\imgcell{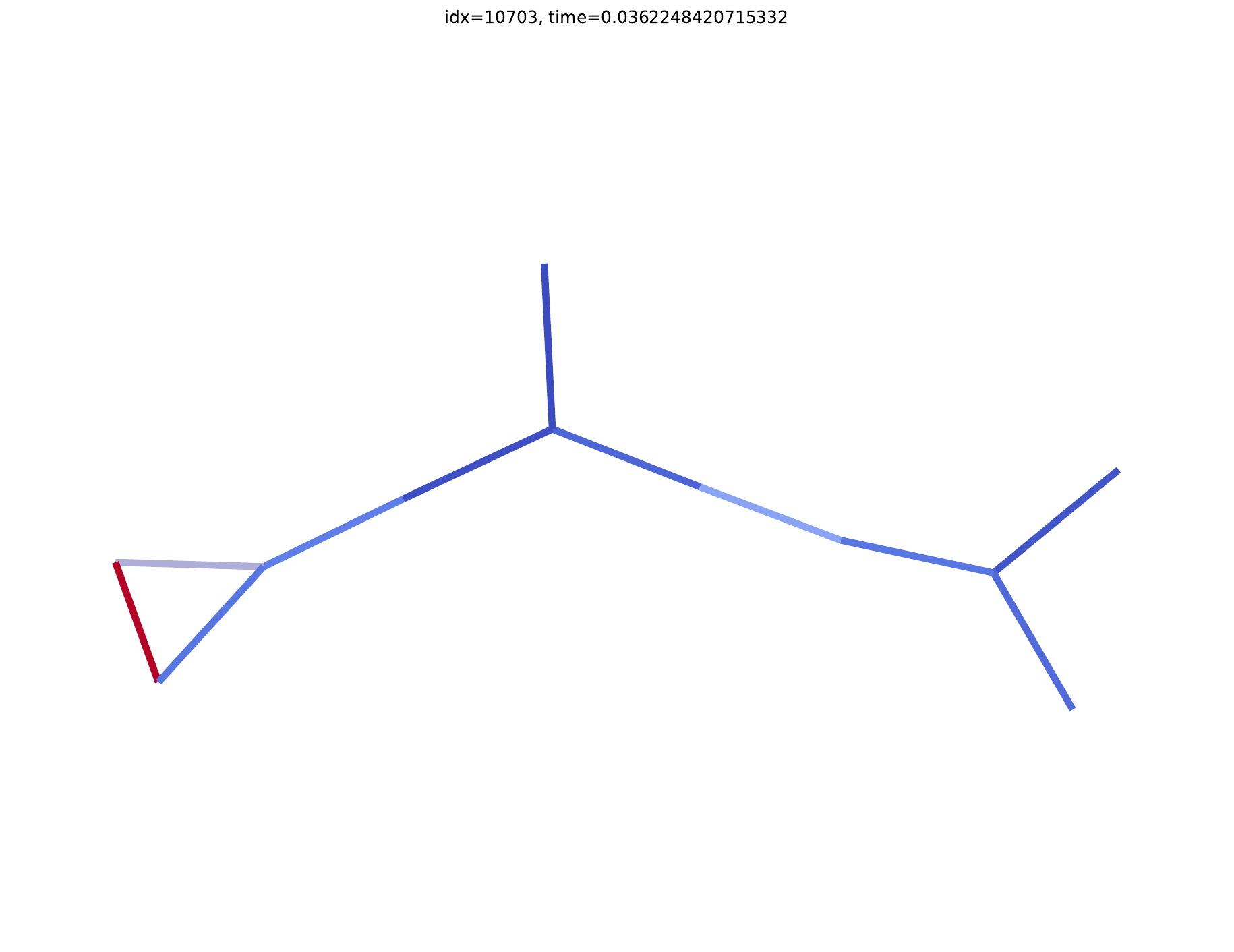} &
\imgcell{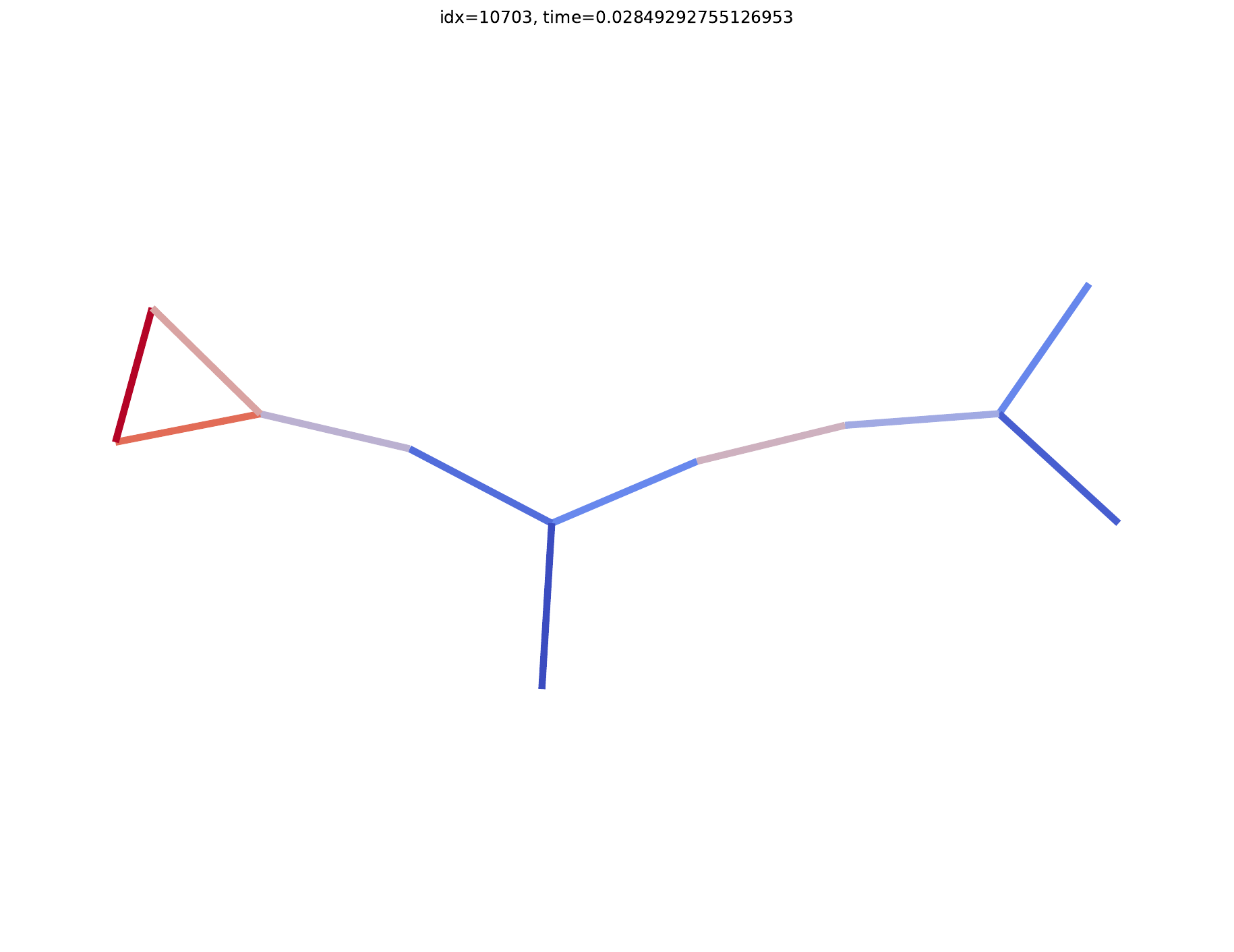} &
\imgcell{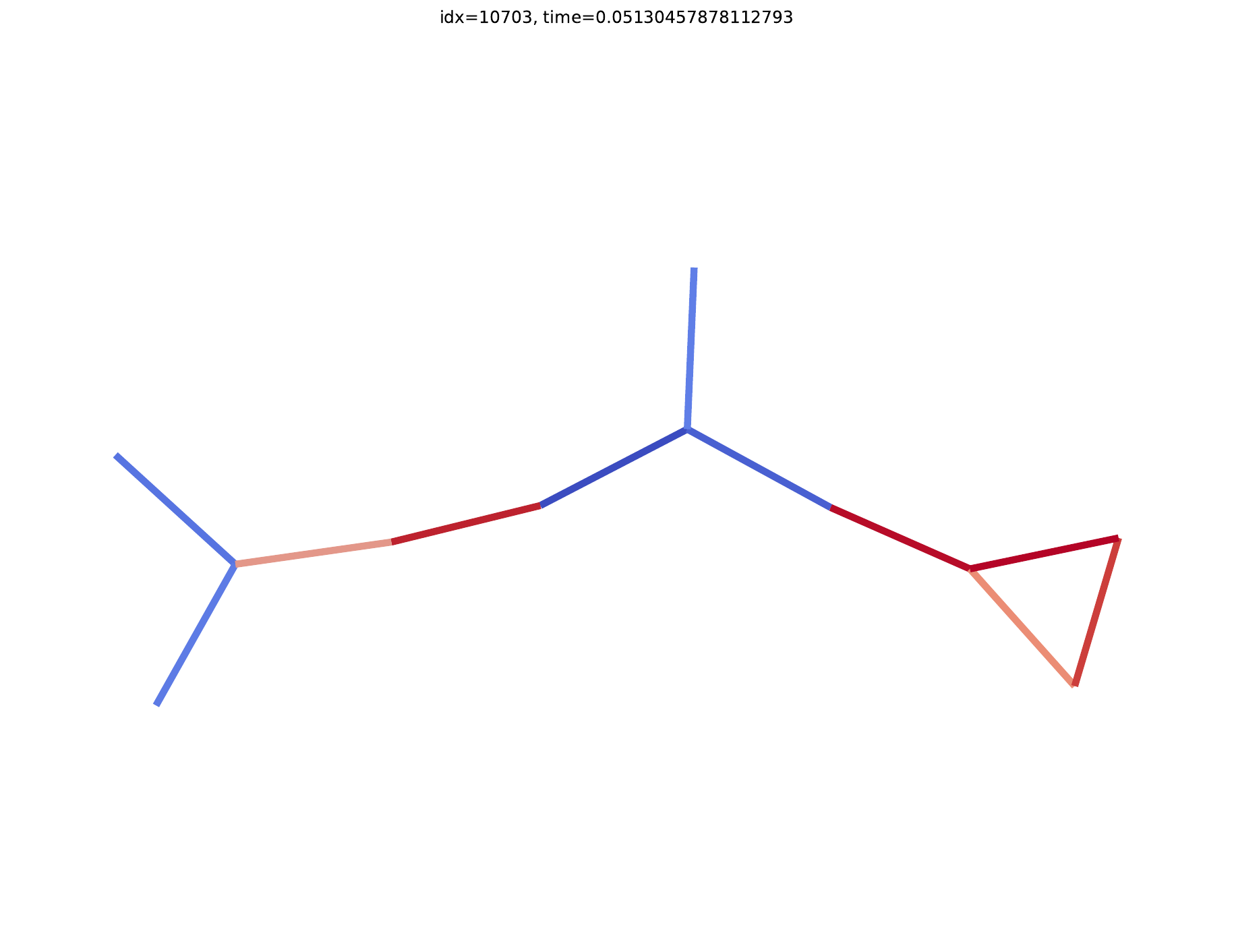} &
\imgcell{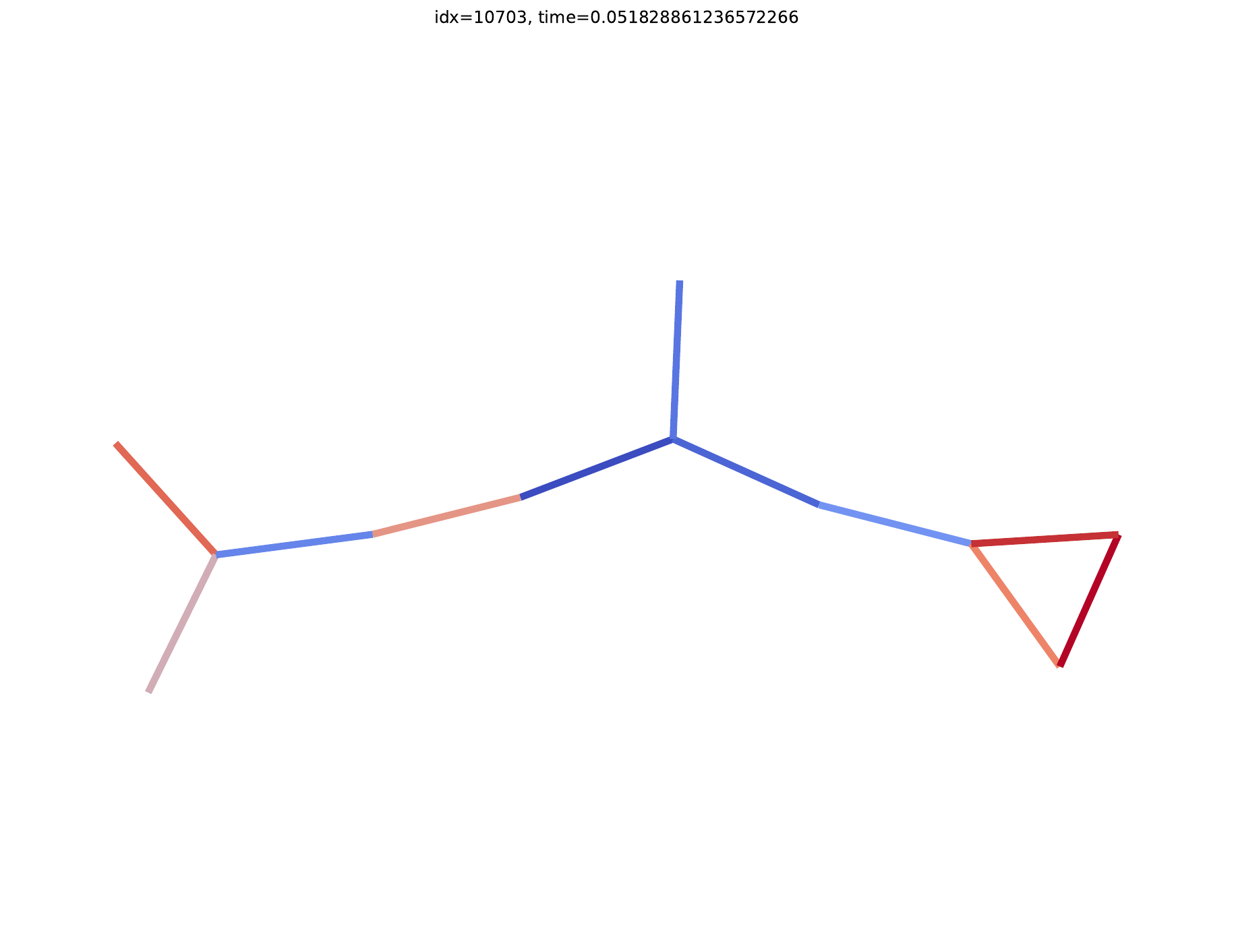} \\

&
t = 0.00s &
t = 0.22s &
t = 0.02s &
t = 0.05s &
t = 74.69s &
t = 0.05s &
t = 0.04s &
t = 0.03s &
t = 0.04s &
t = 0.03s &
t = 0.05s &
t = 0.05s \\

\makecell{\bfseries grafo8329.78\\N = 34\\M = 38} &
\imgcell{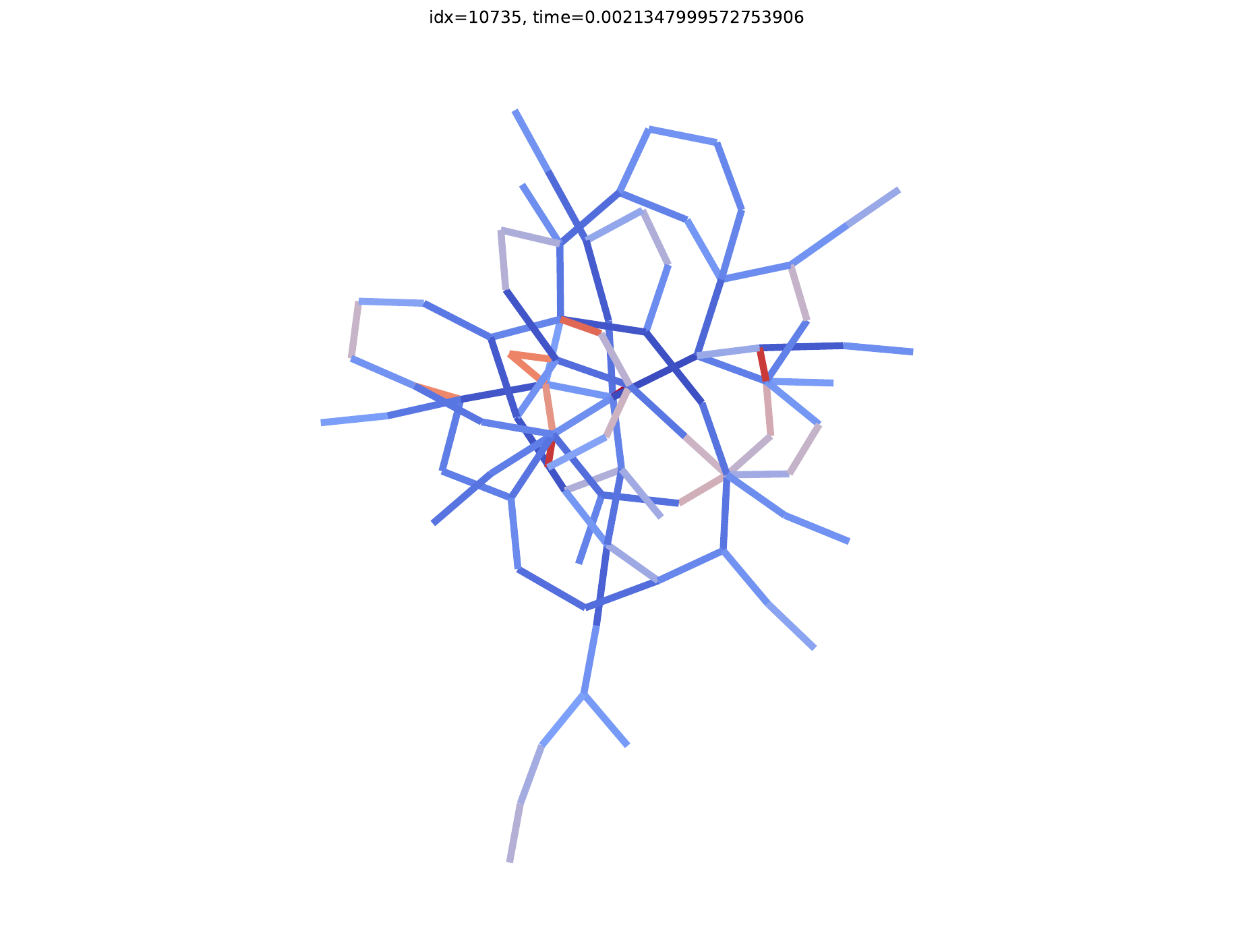} &
\imgcell{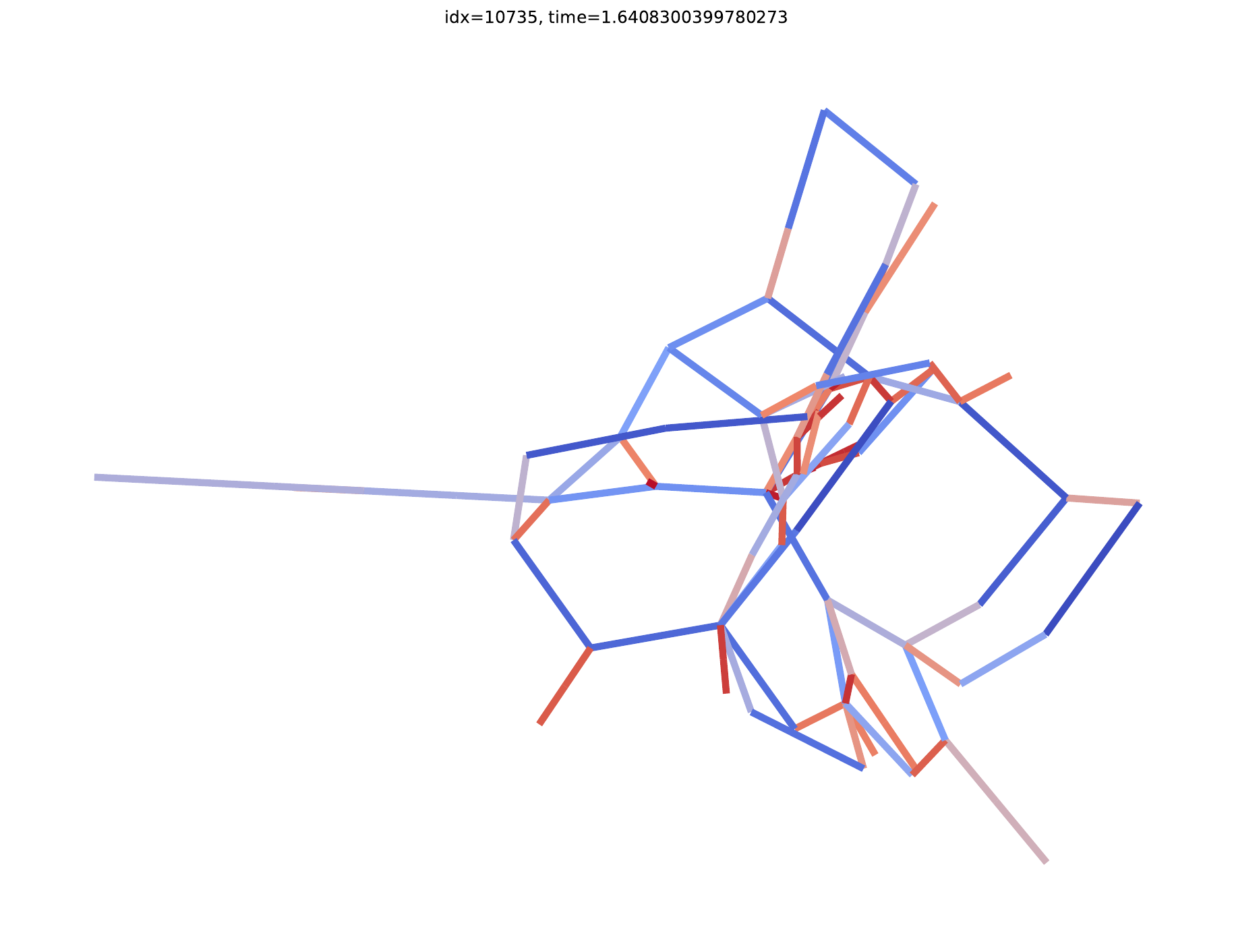} &
\imgcell{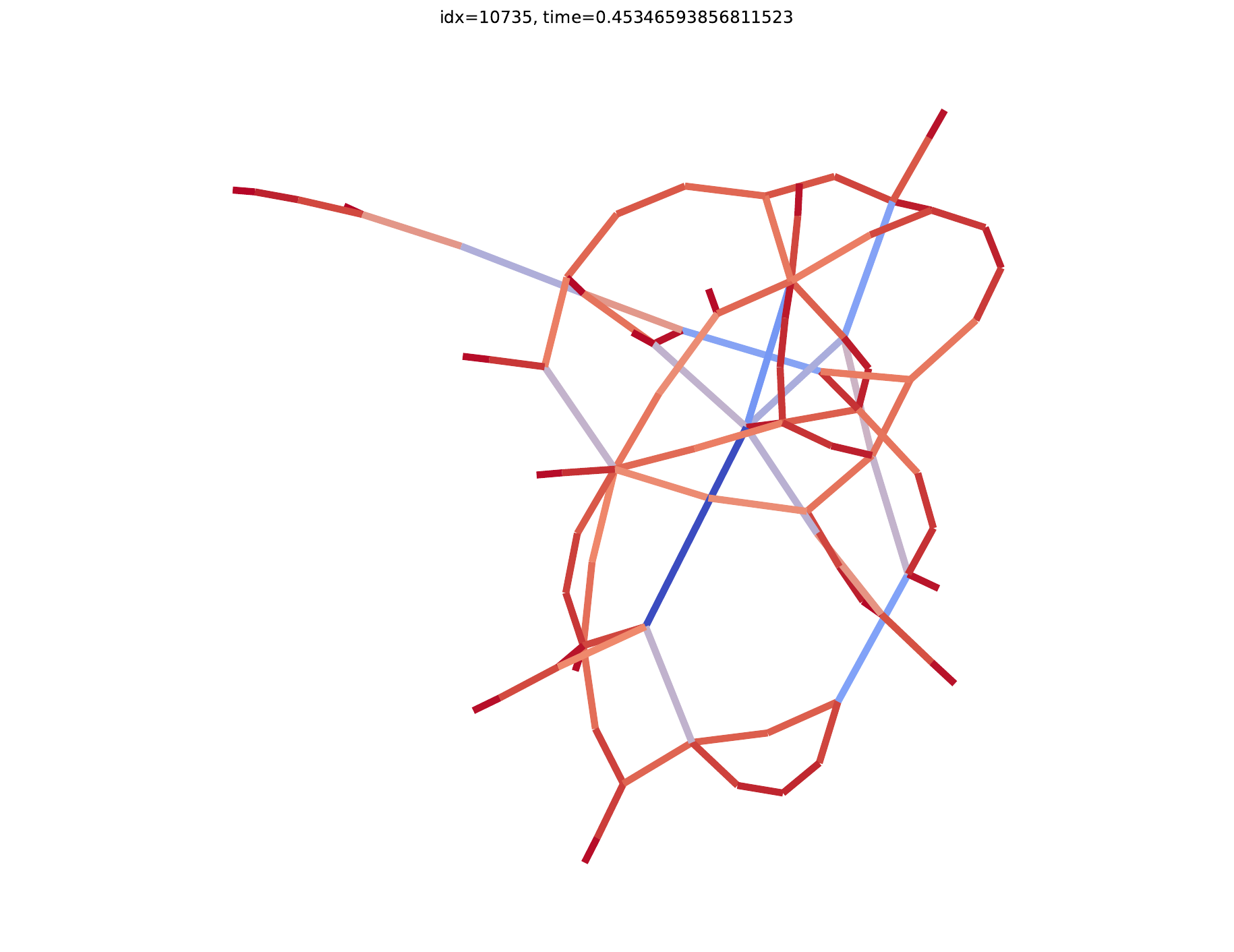} &
\imgcell{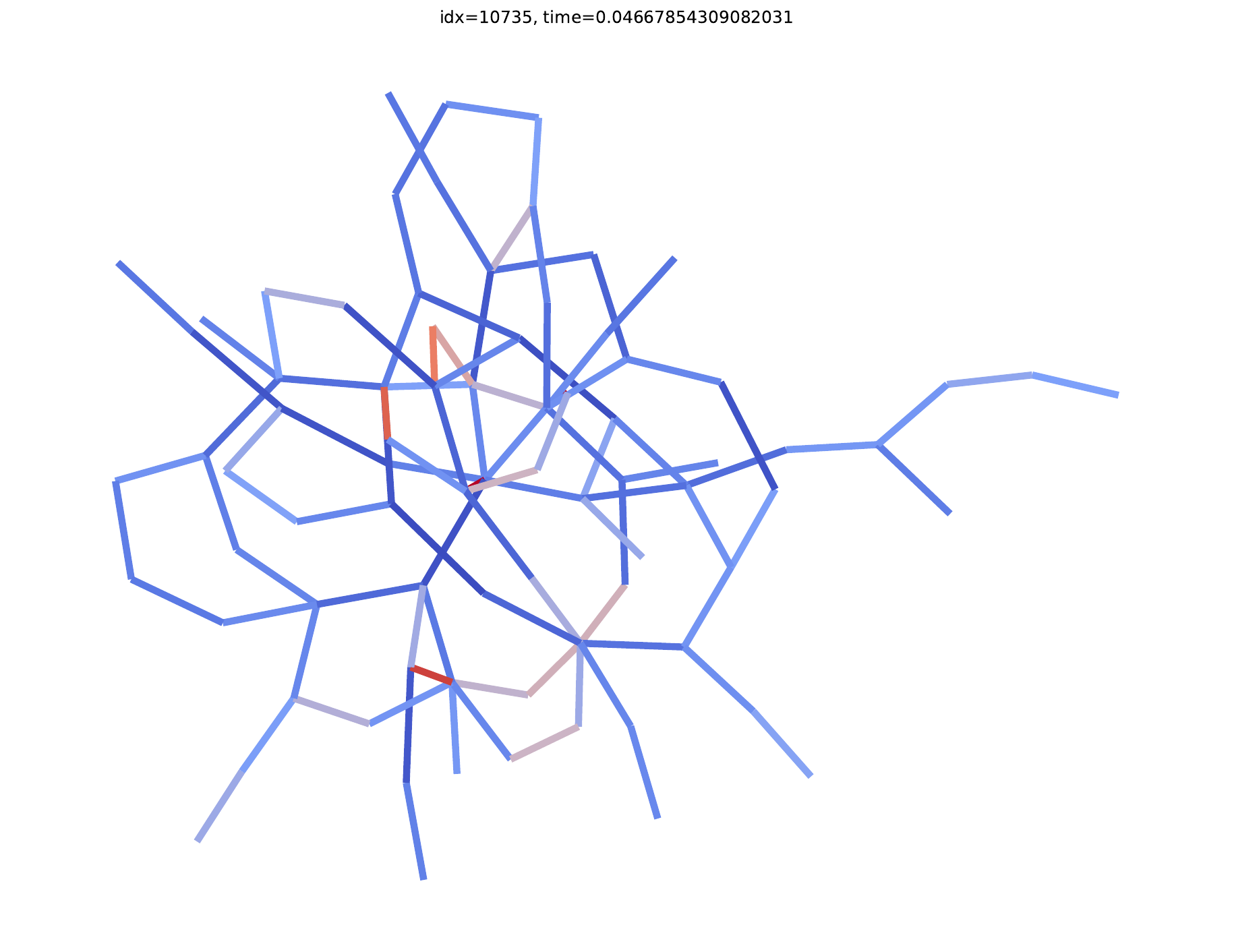} &
\imgcell{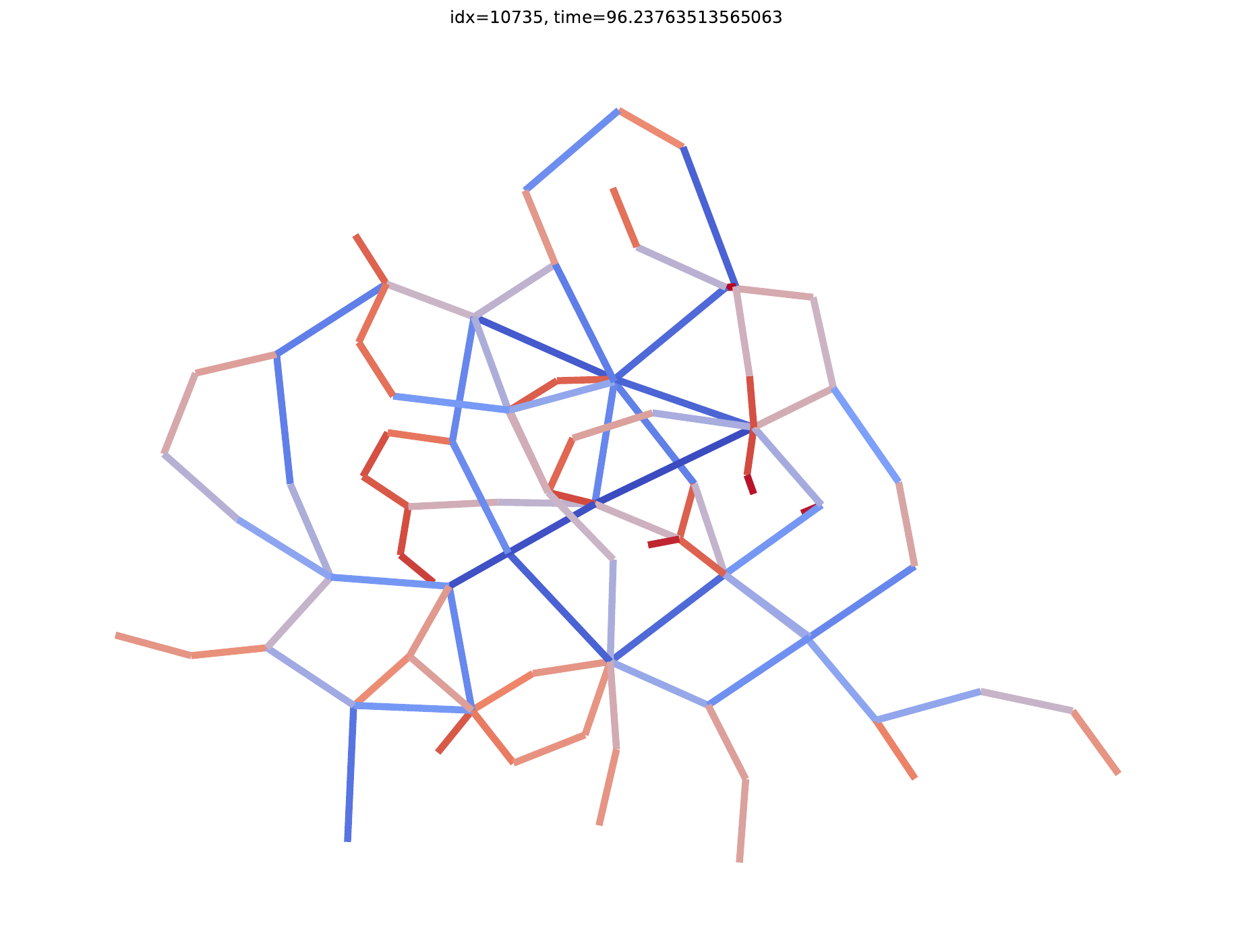} &
\imgcell{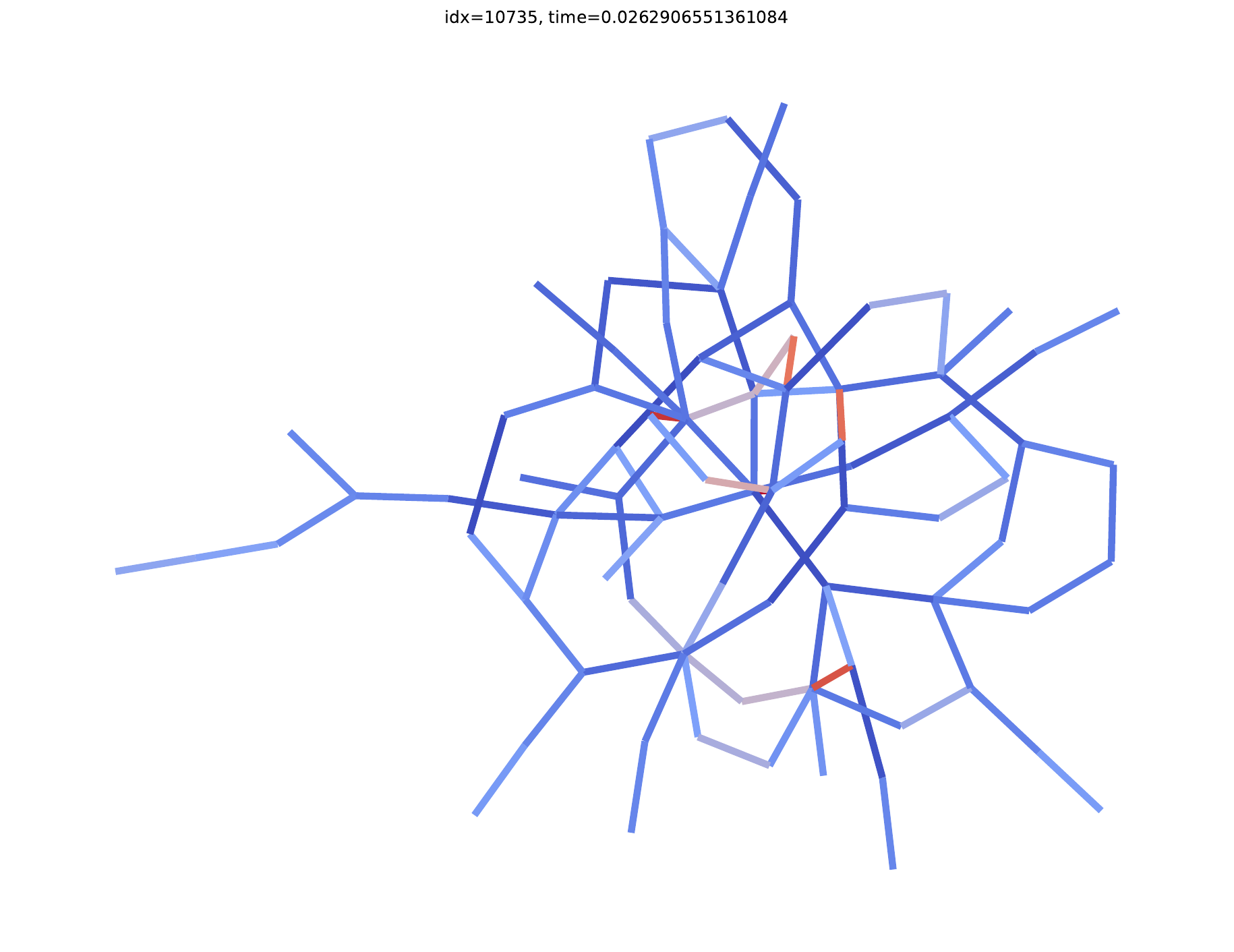} &
\imgcell{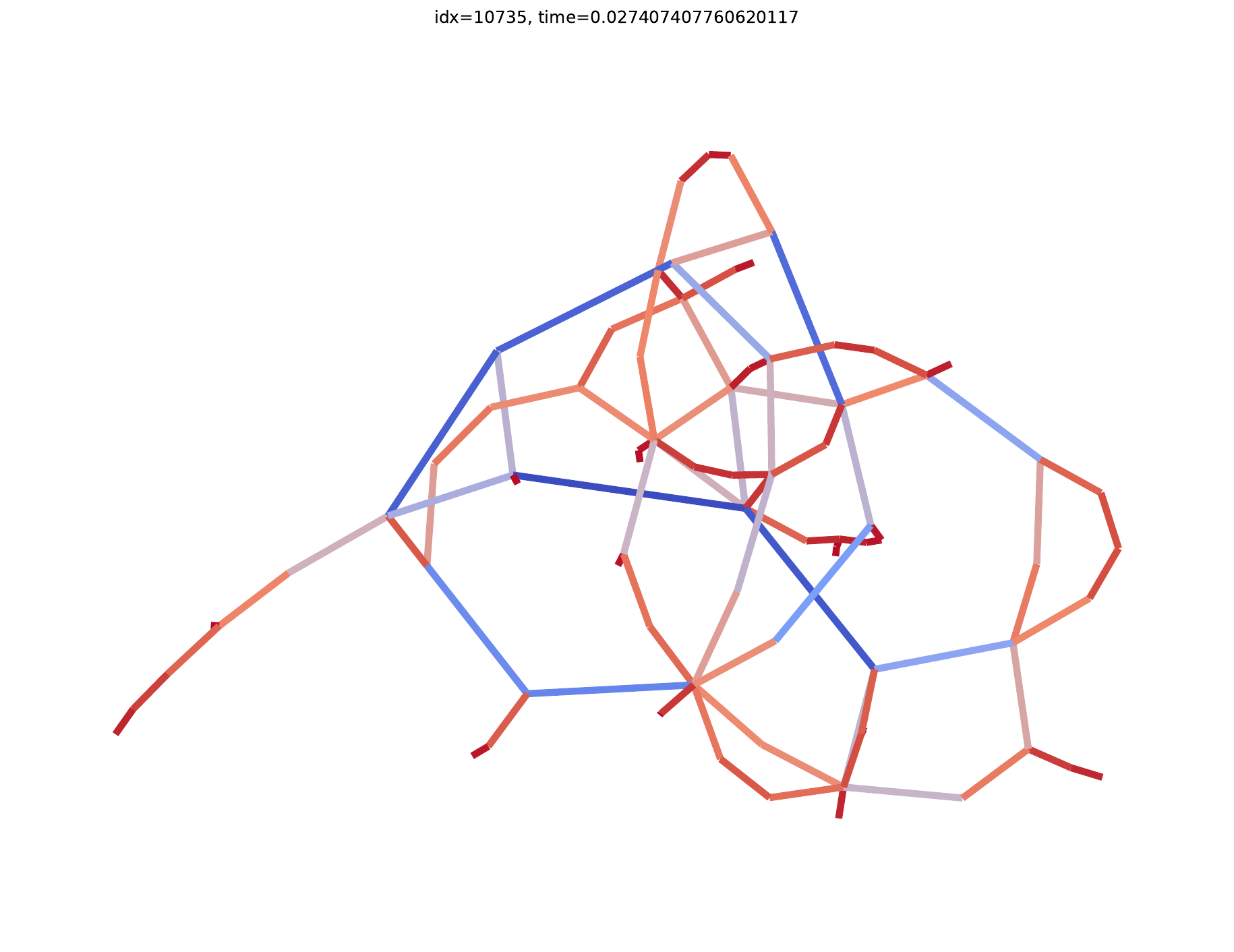} &
\imgcell{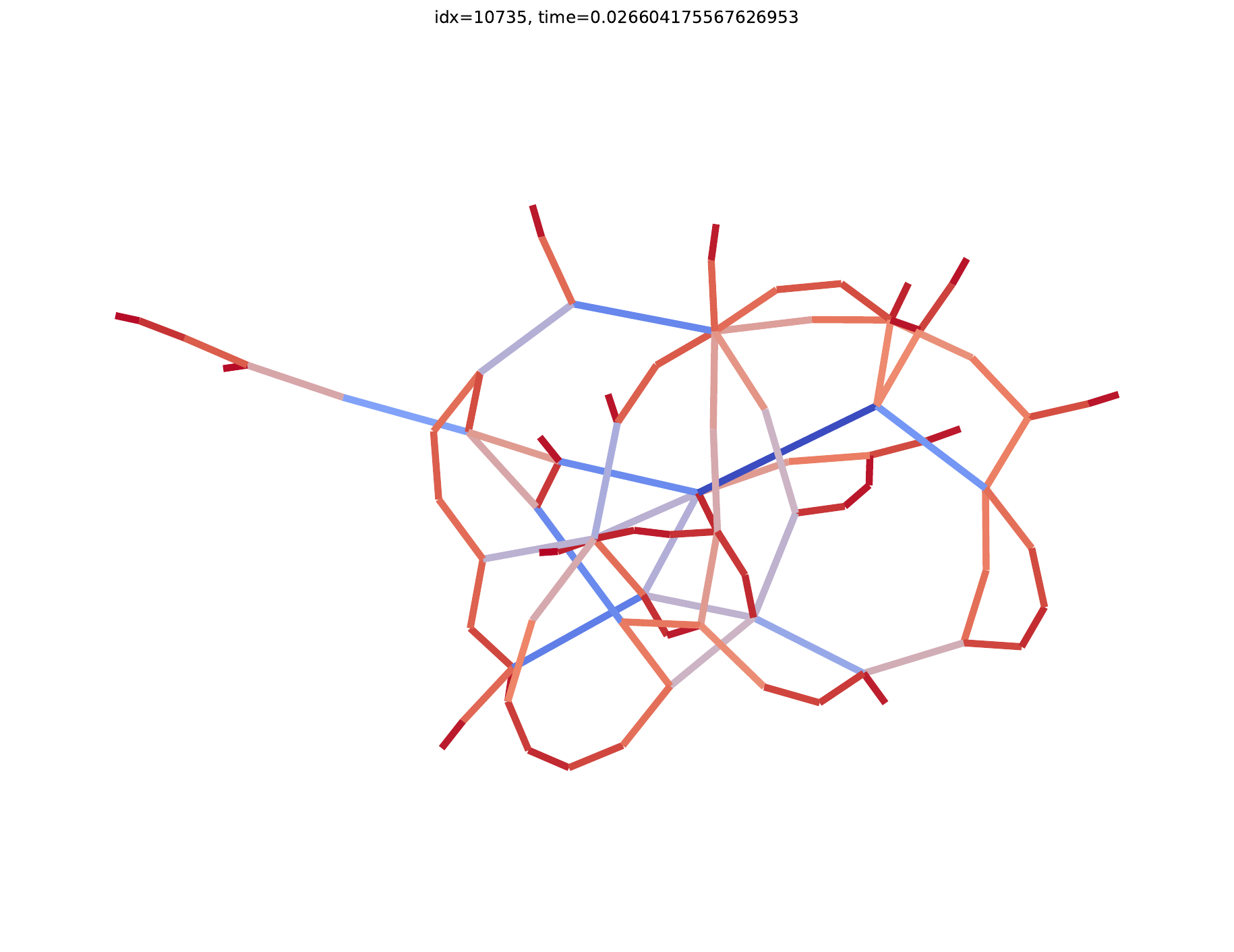} &
\imgcell{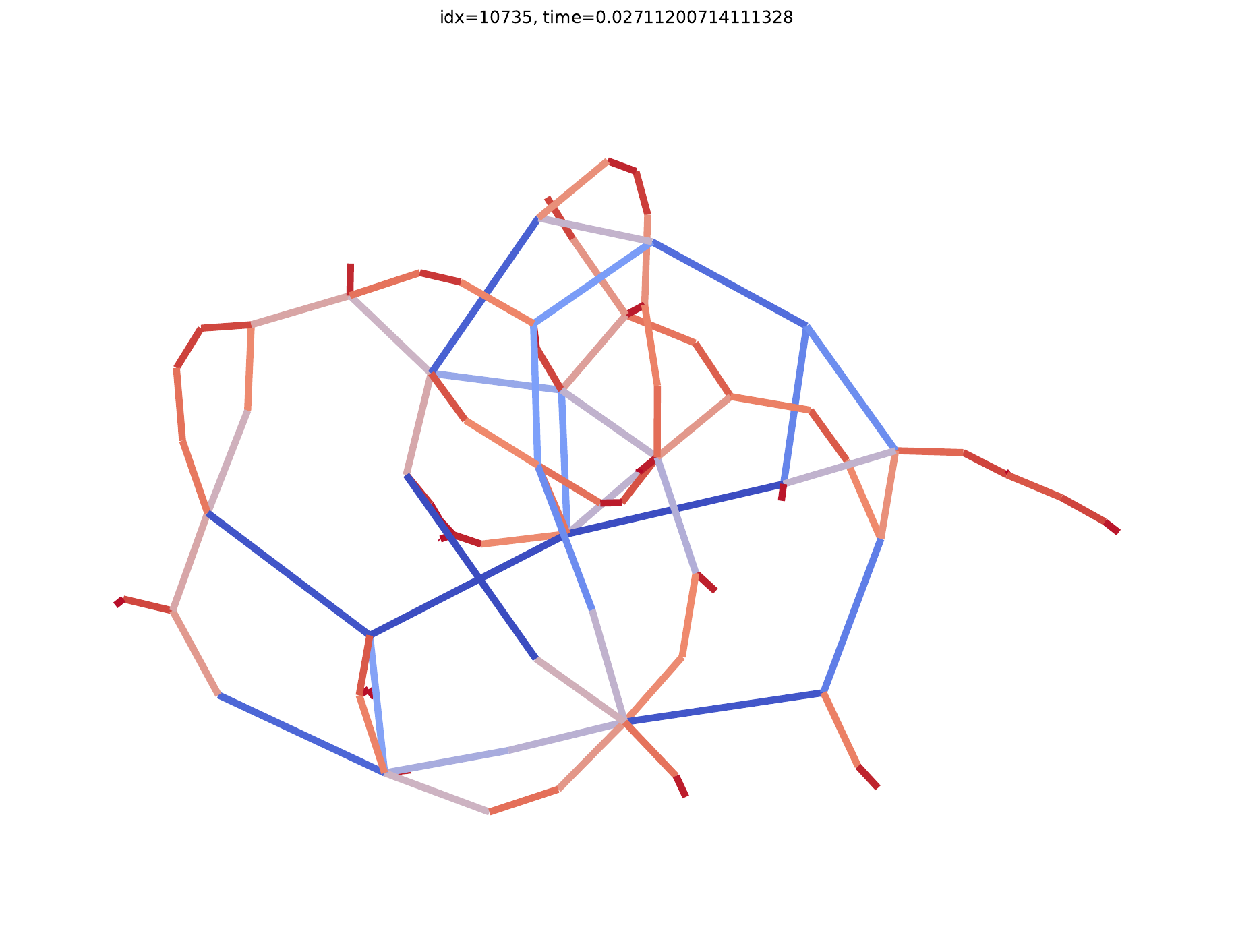} &
\imgcell{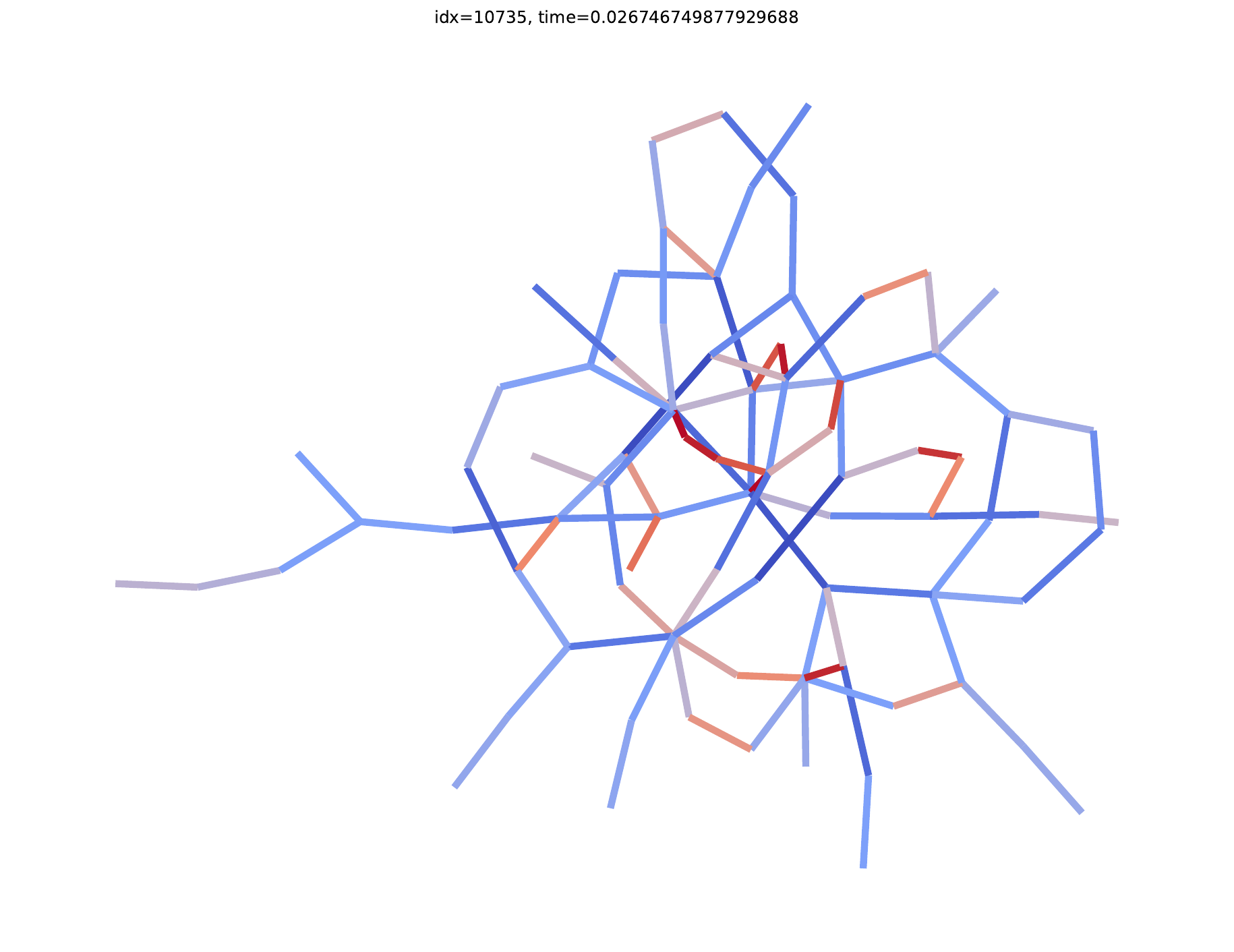} &
\imgcell{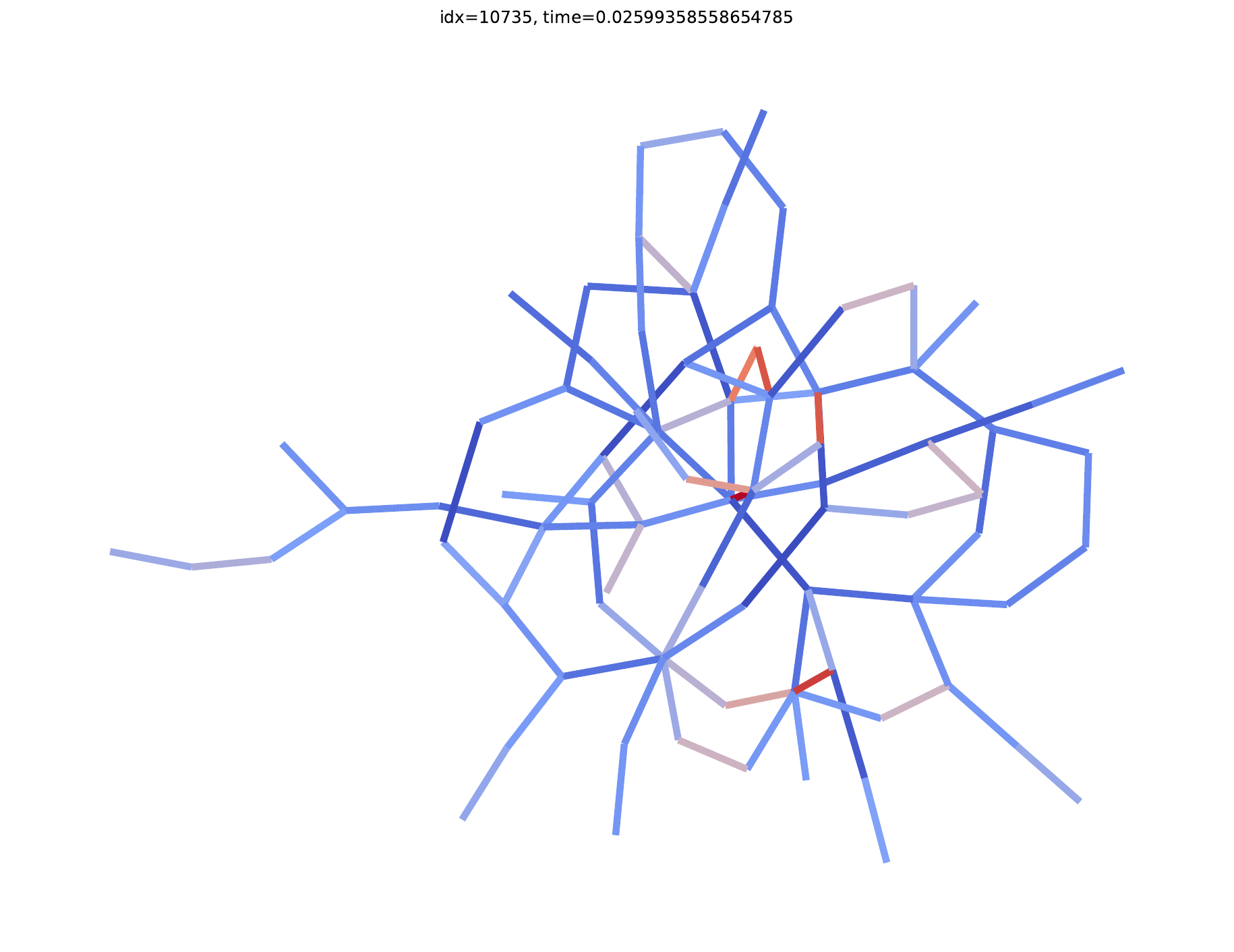} &
\imgcell{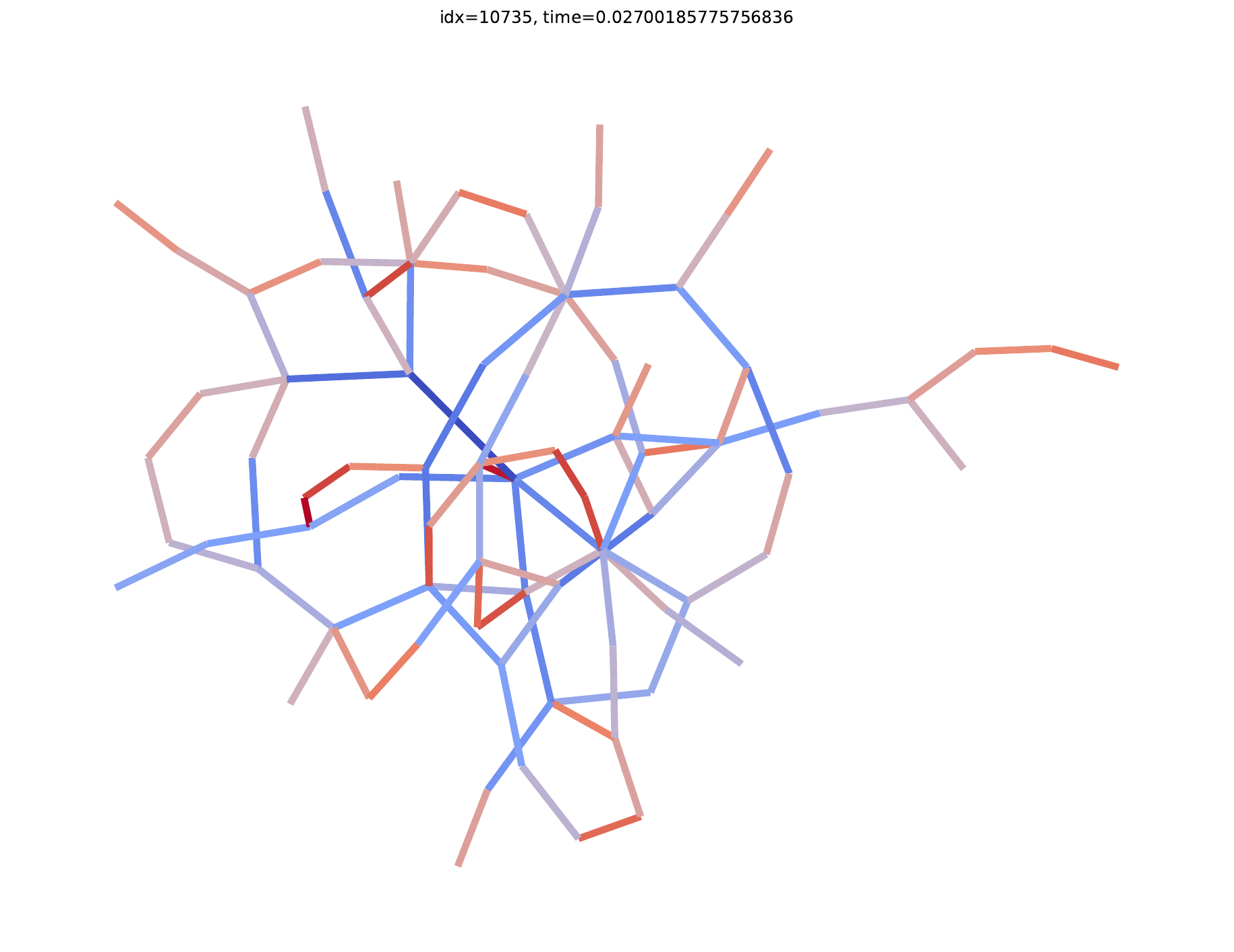} \\

&
t = 0.00s &
t = 1.64s &
t = 0.45s &
t = 0.05s &
t = 96.24s &
t = 0.03s &
t = 0.03s &
t = 0.04s &
t = 0.03s &
t = 0.03s &
t = 0.03s &
t = 0.03s \\

\makecell{\bfseries grafo10553.98\\N = 85\\M = 123} &
\imgcell{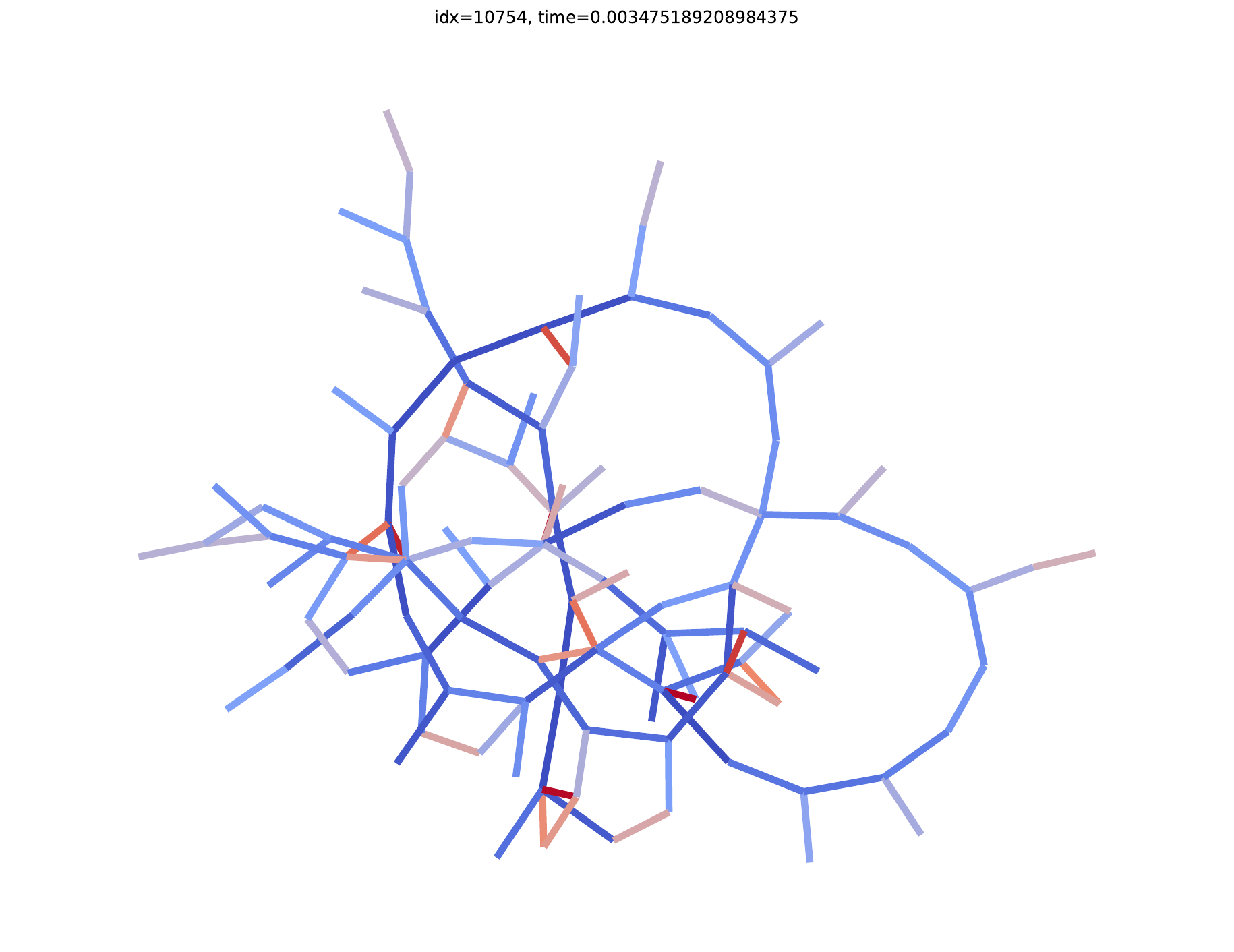} &
\imgcell{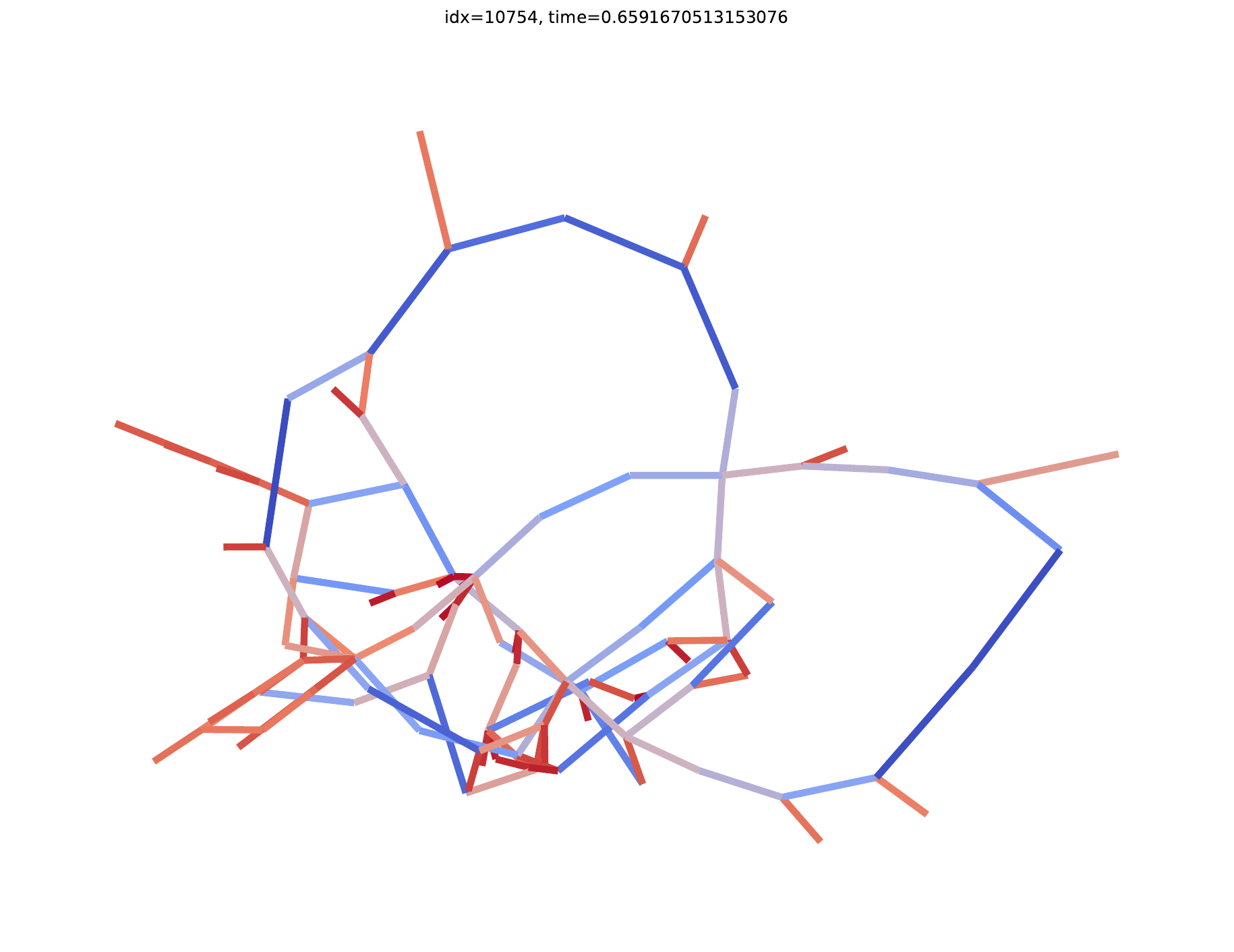} &
\imgcell{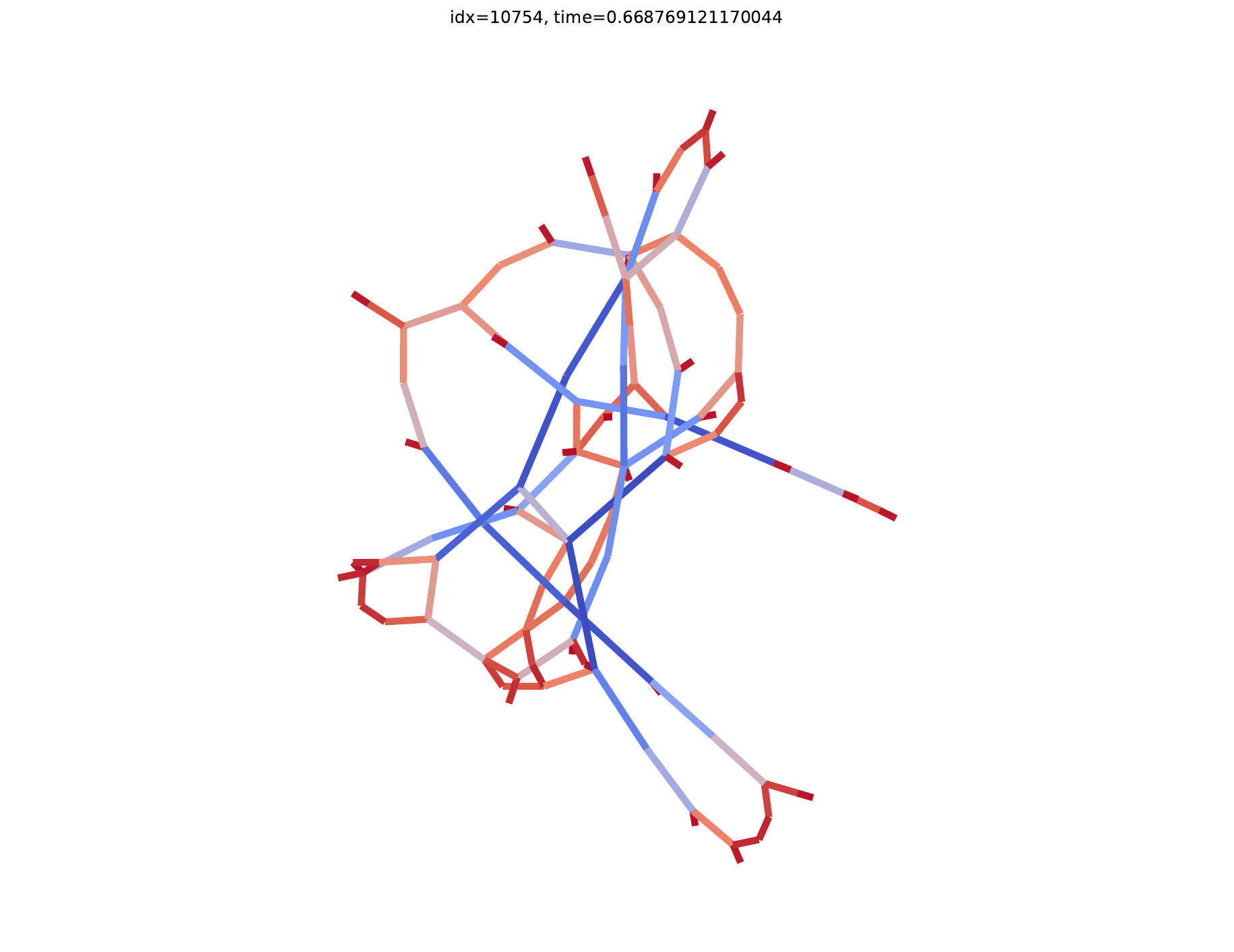} &
\imgcell{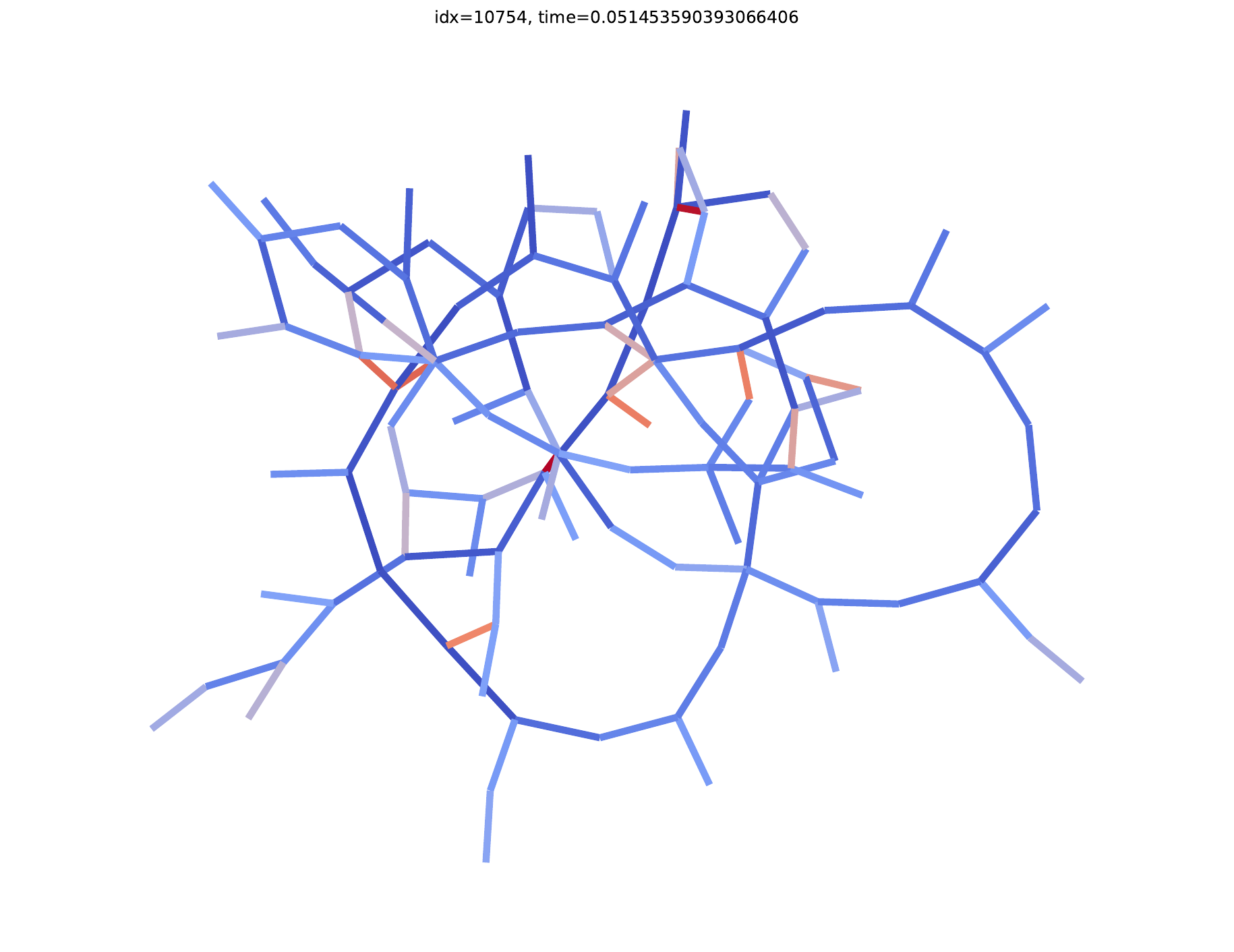} &
\imgcell{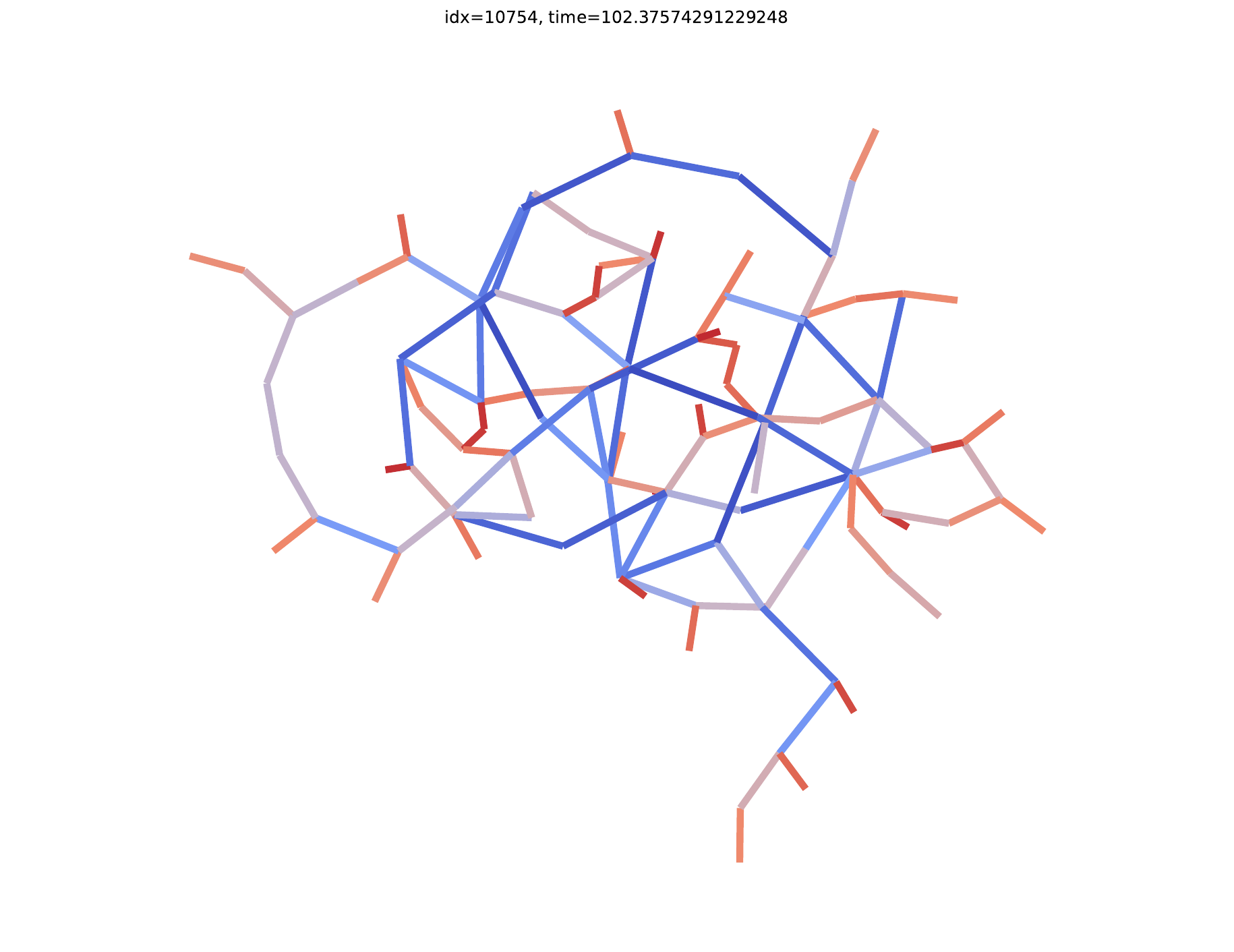} &
\imgcell{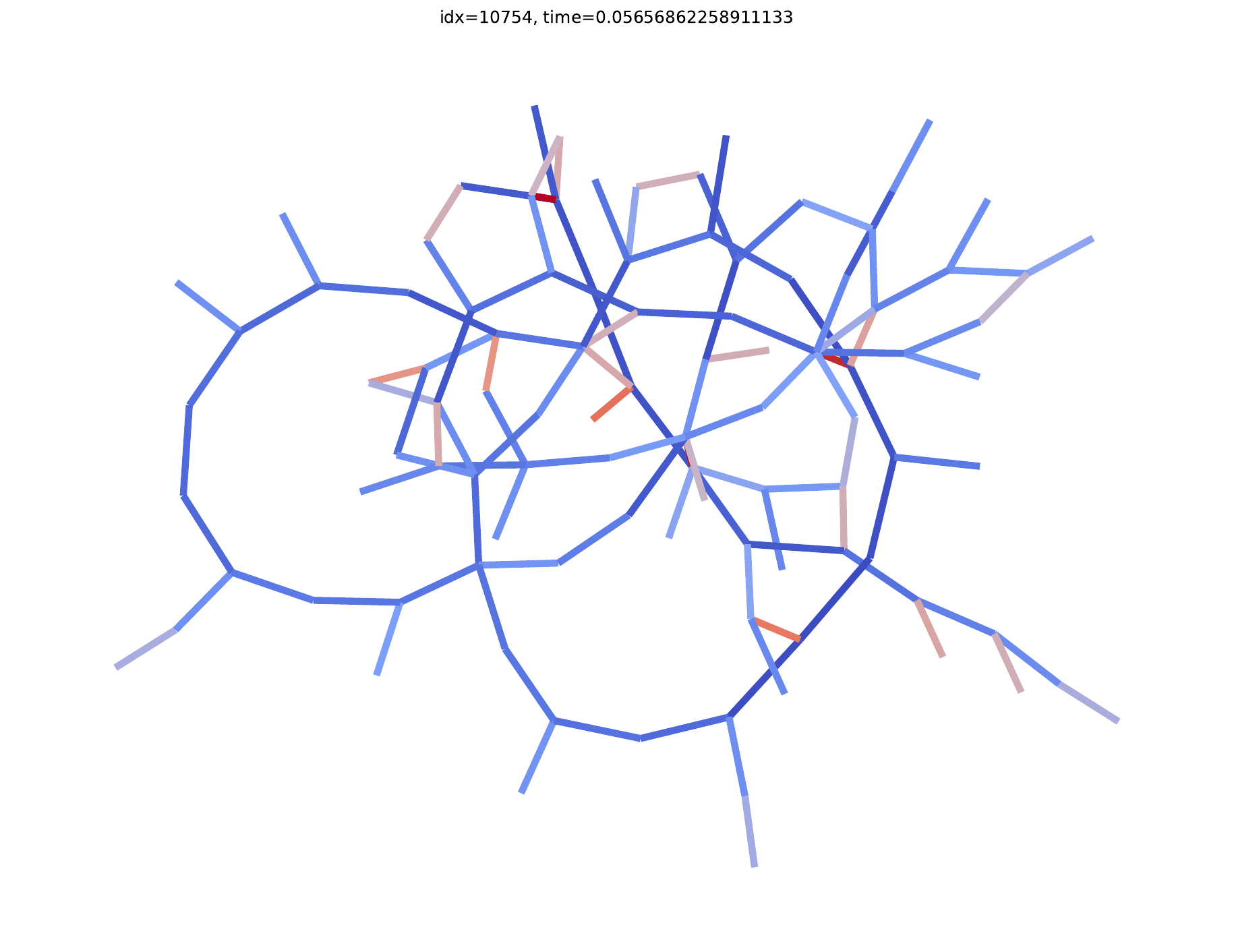} &
\imgcell{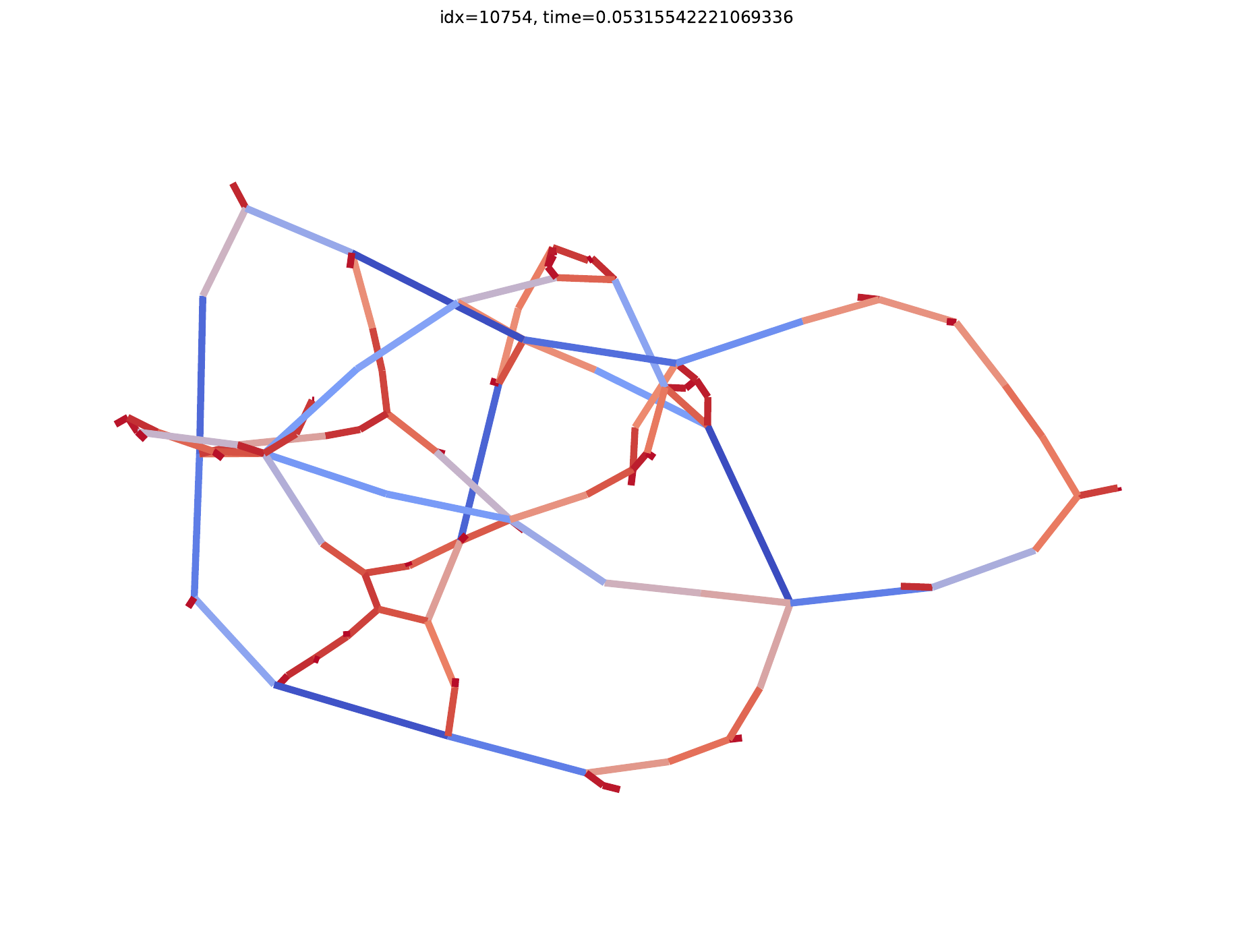} &
\imgcell{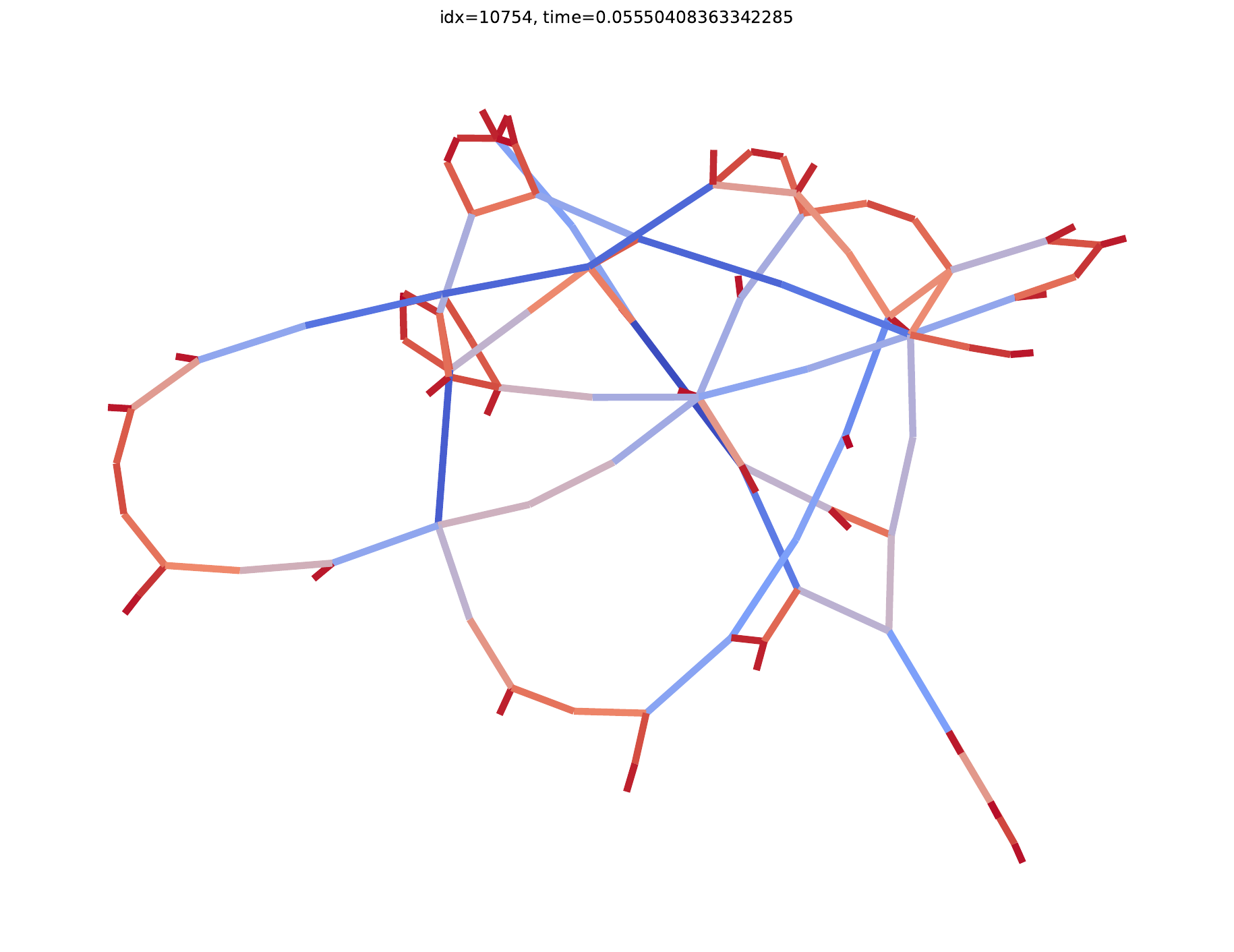} &
\imgcell{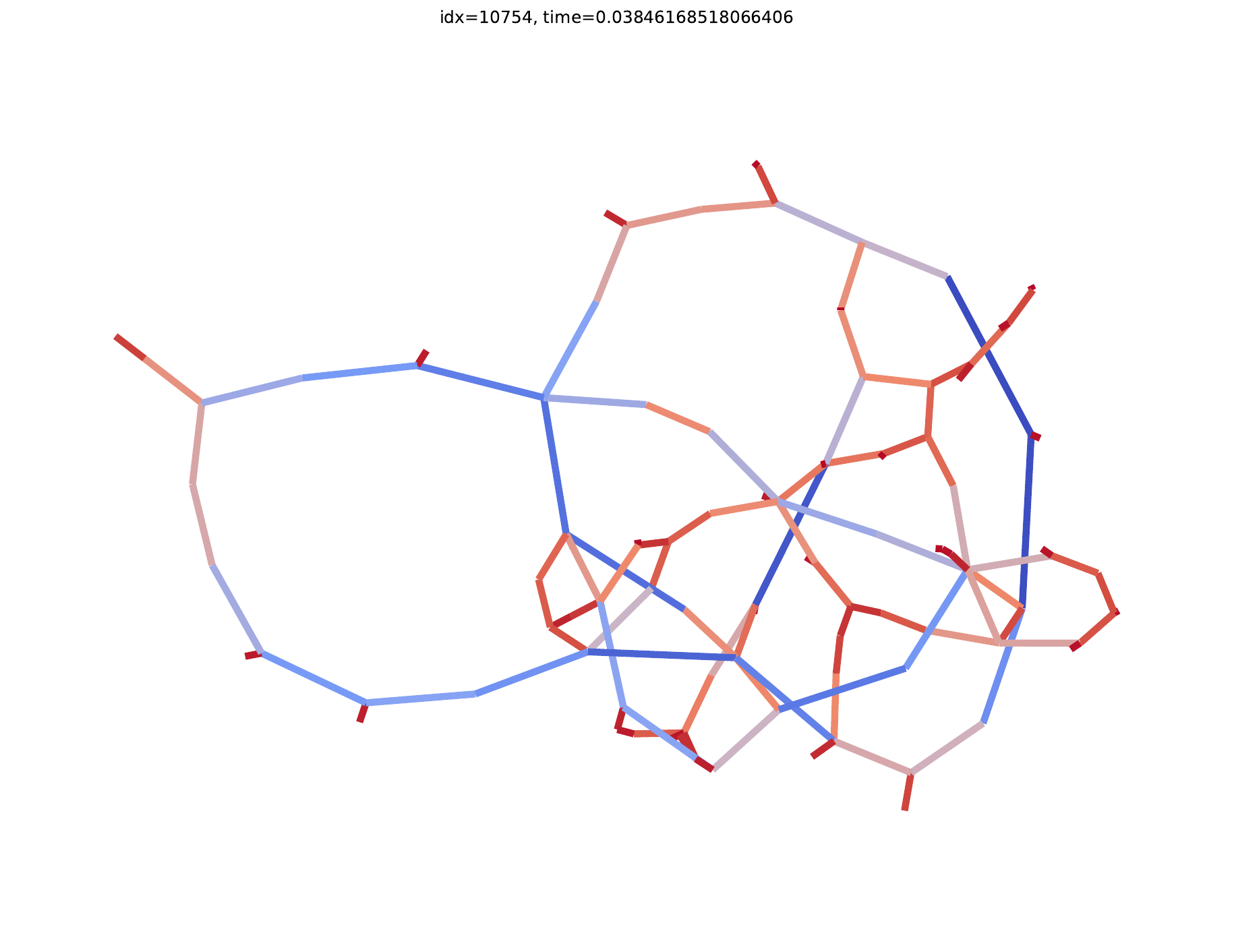} &
\imgcell{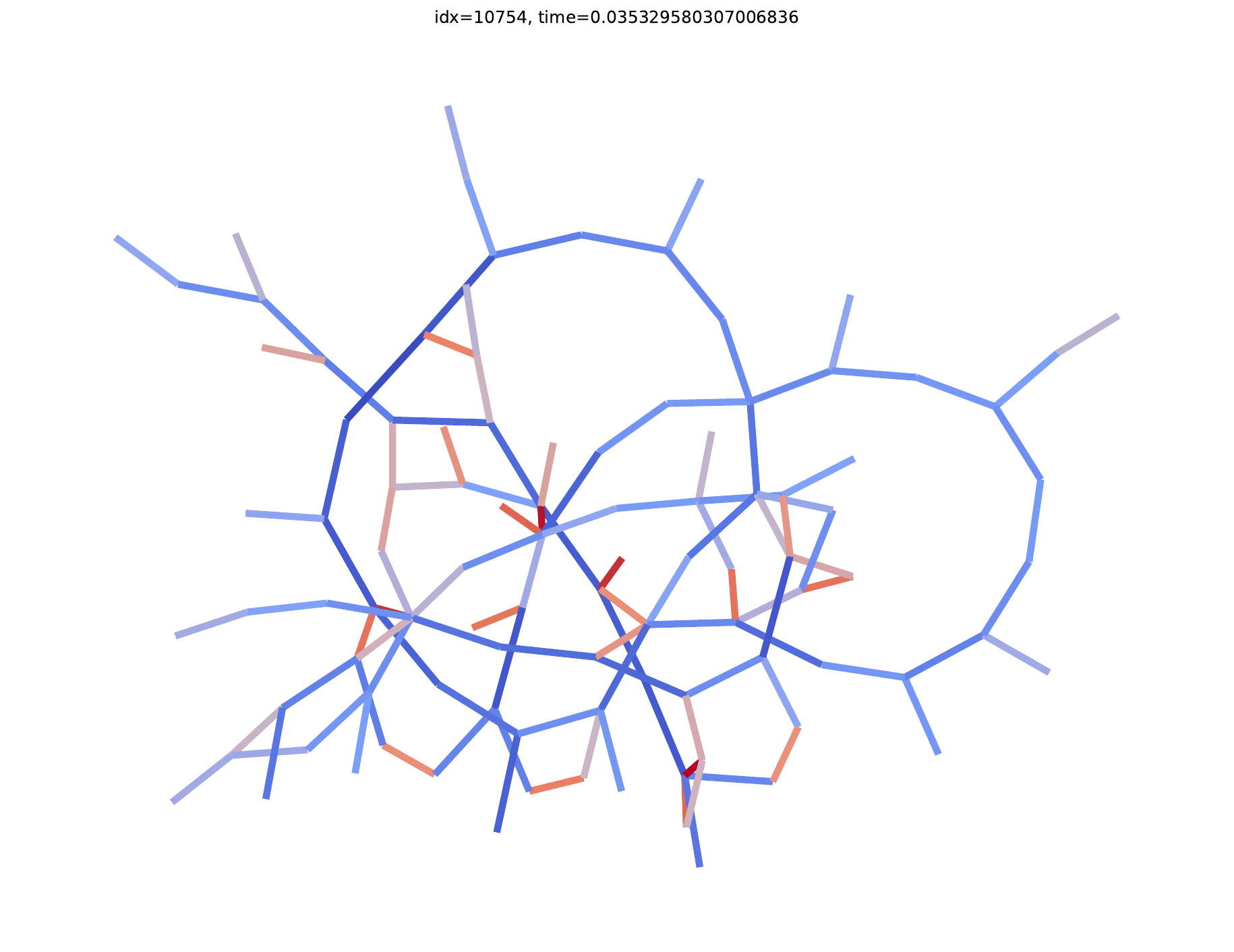} &
\imgcell{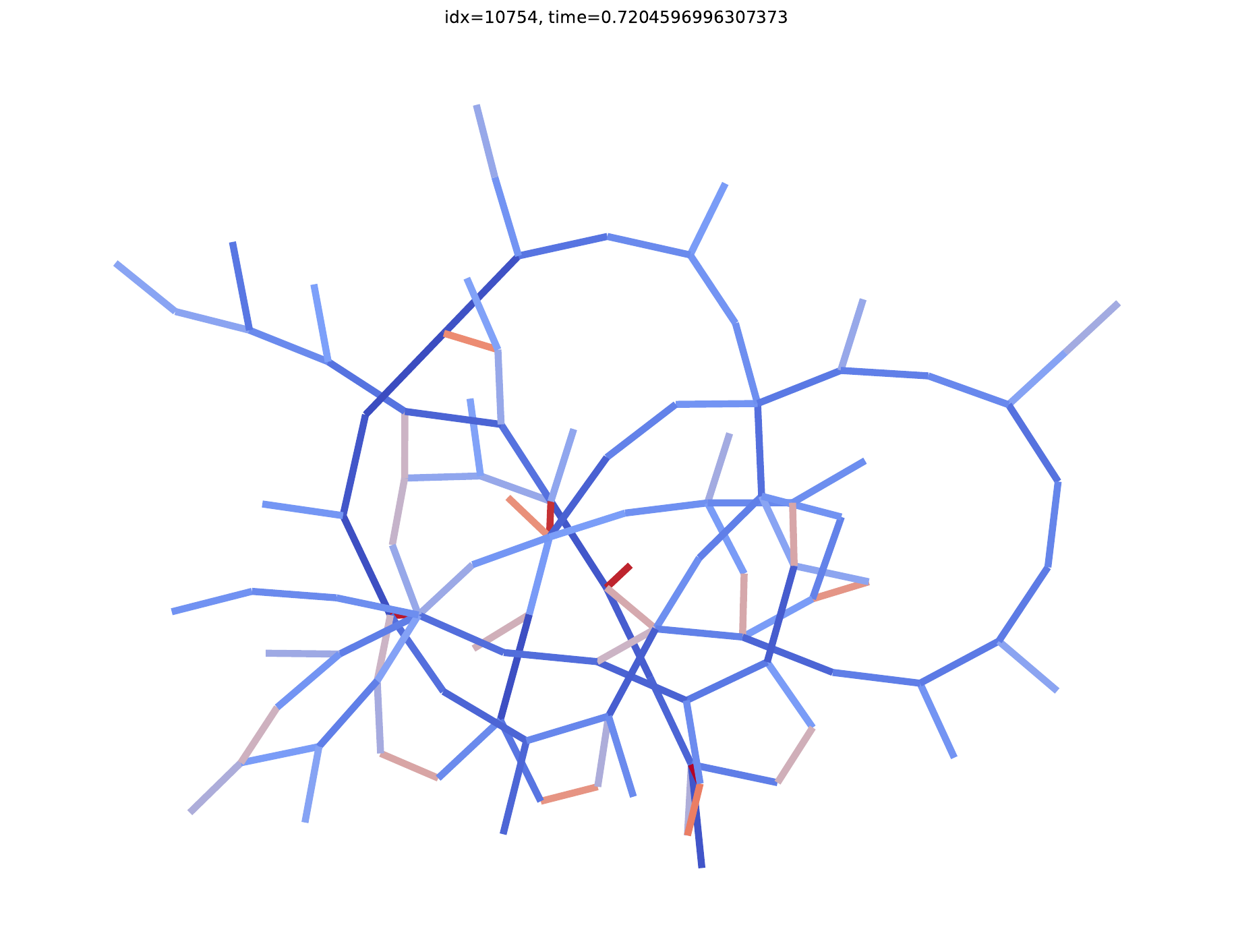} &
\imgcell{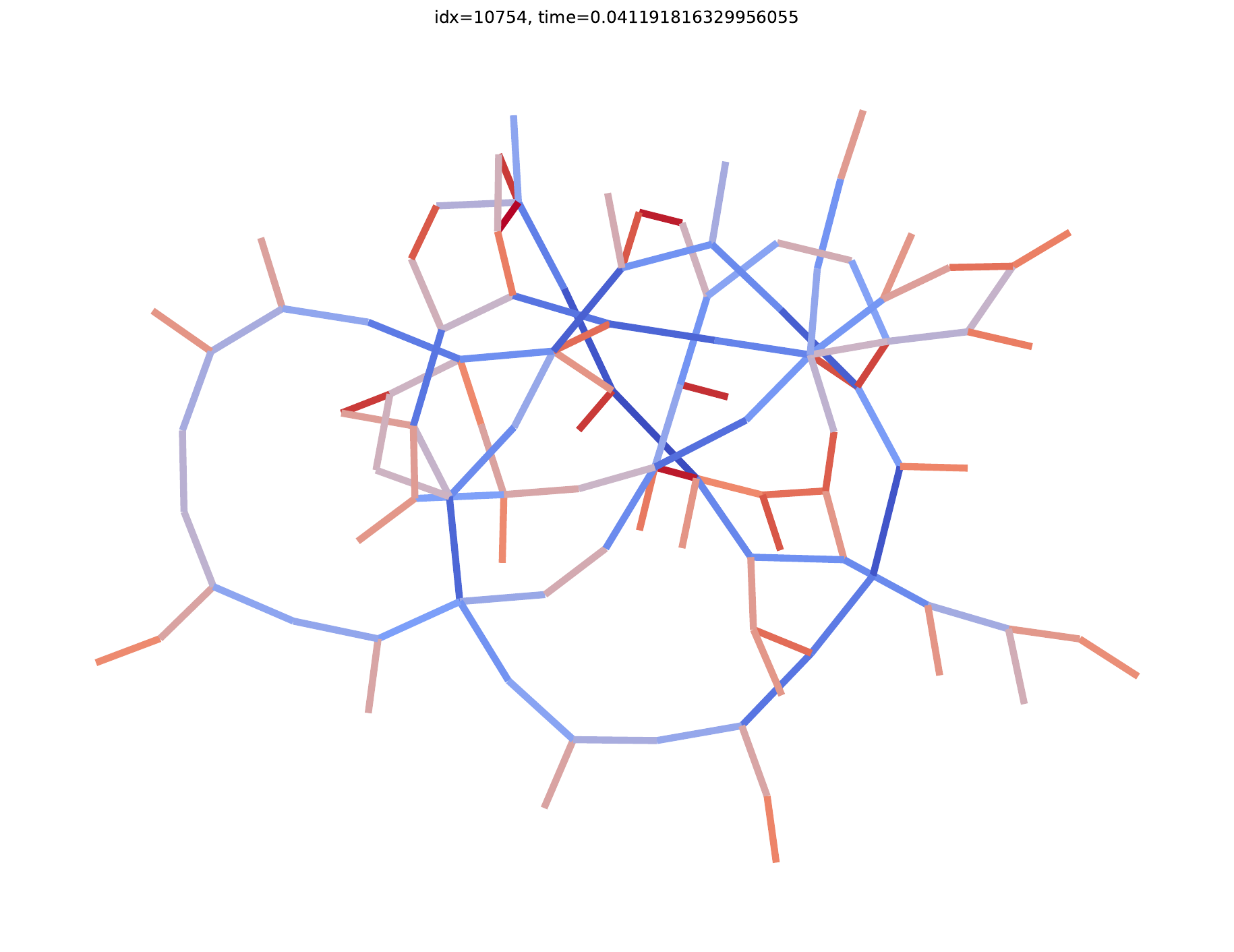} \\

&
t = 0.00s &
t = 0.66s &
t = 0.67s &
t = 0.05s &
t = 102.38s &
t = 0.06s &
t = 0.05s &
t = 0.04s &
t = 0.04s &
t = 0.04s &
t = 0.05s &
t = 0.04s \\

\makecell{\bfseries grafo597.17\\N = 70\\M = 91} &
\imgcell{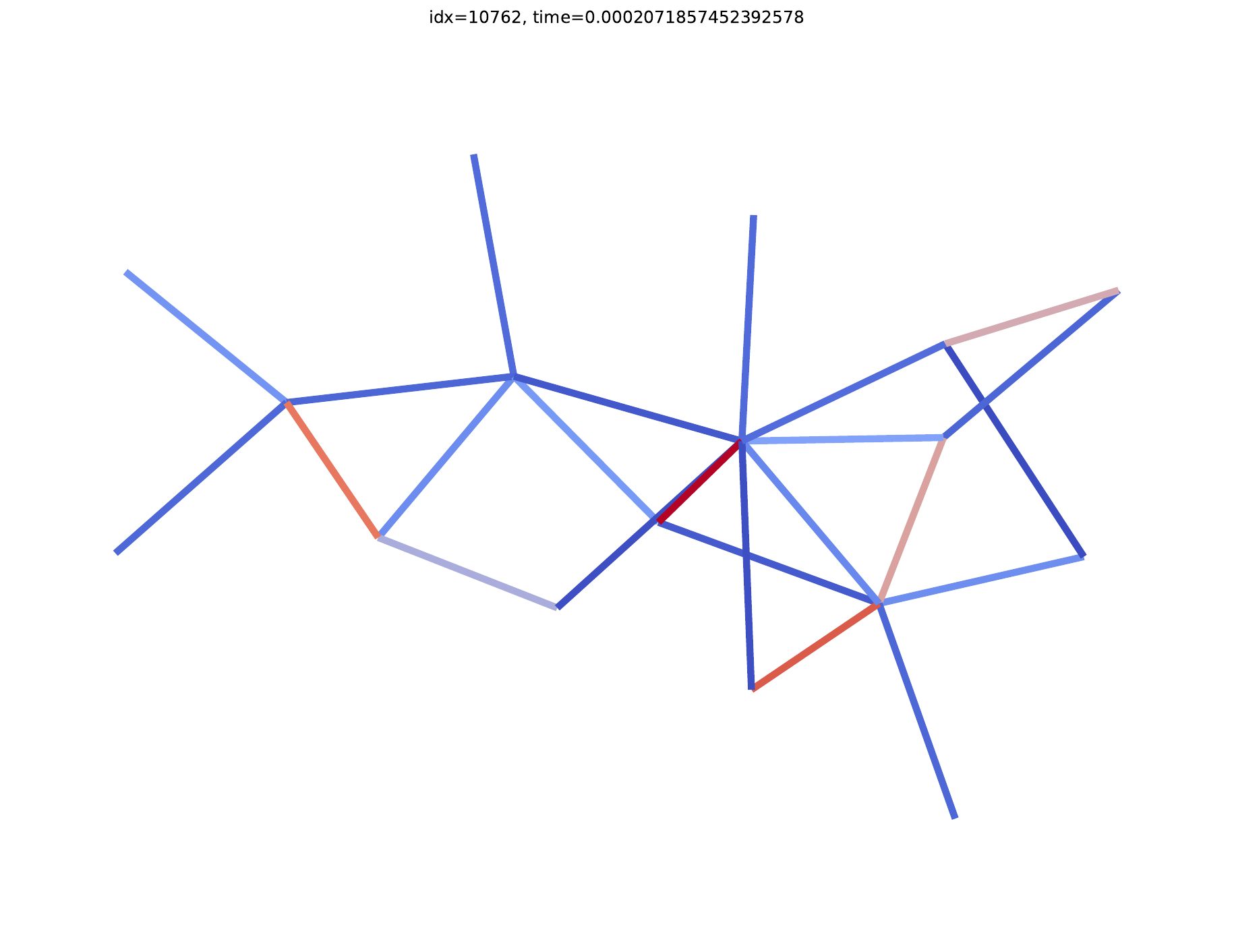} &
\imgcell{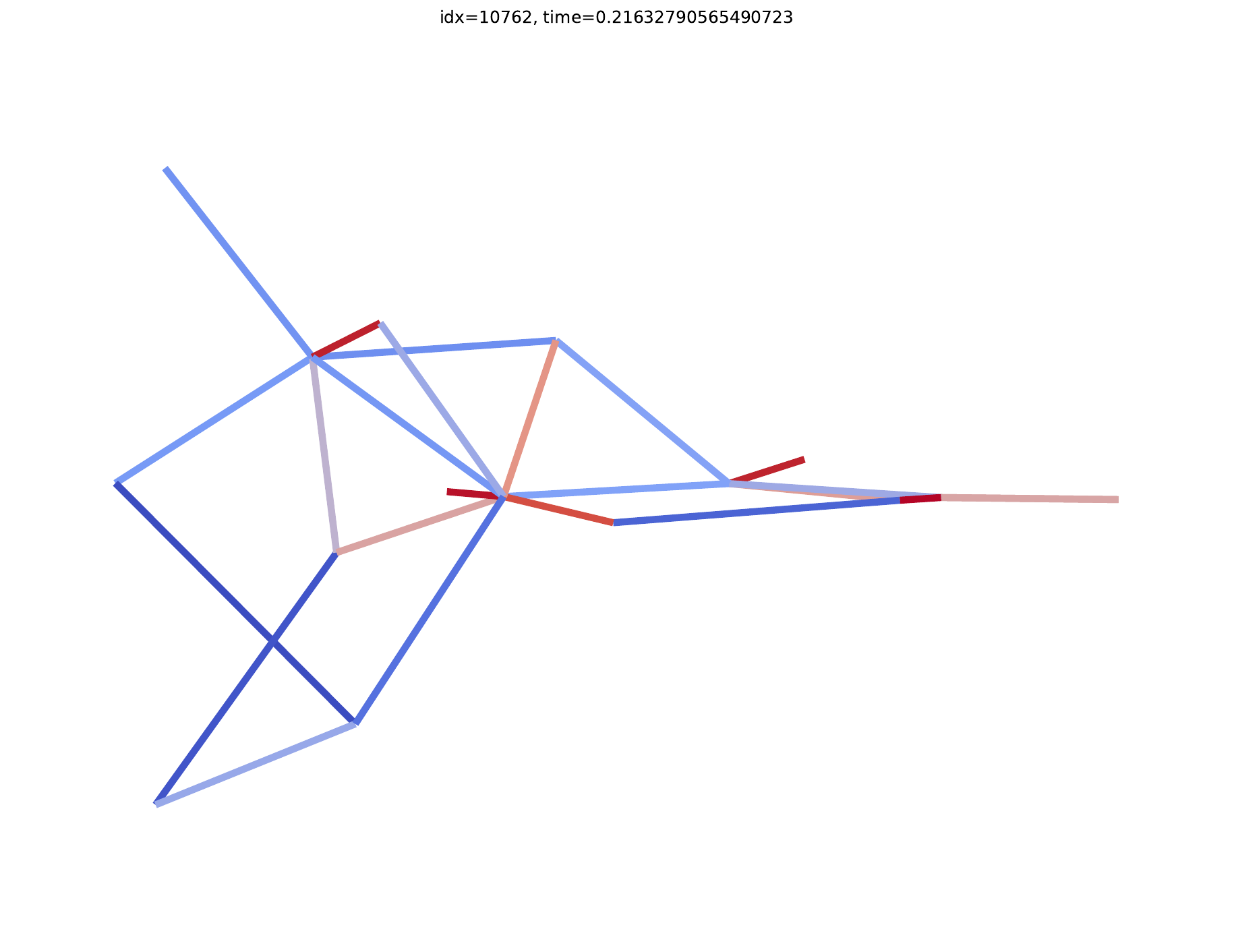} &
\imgcell{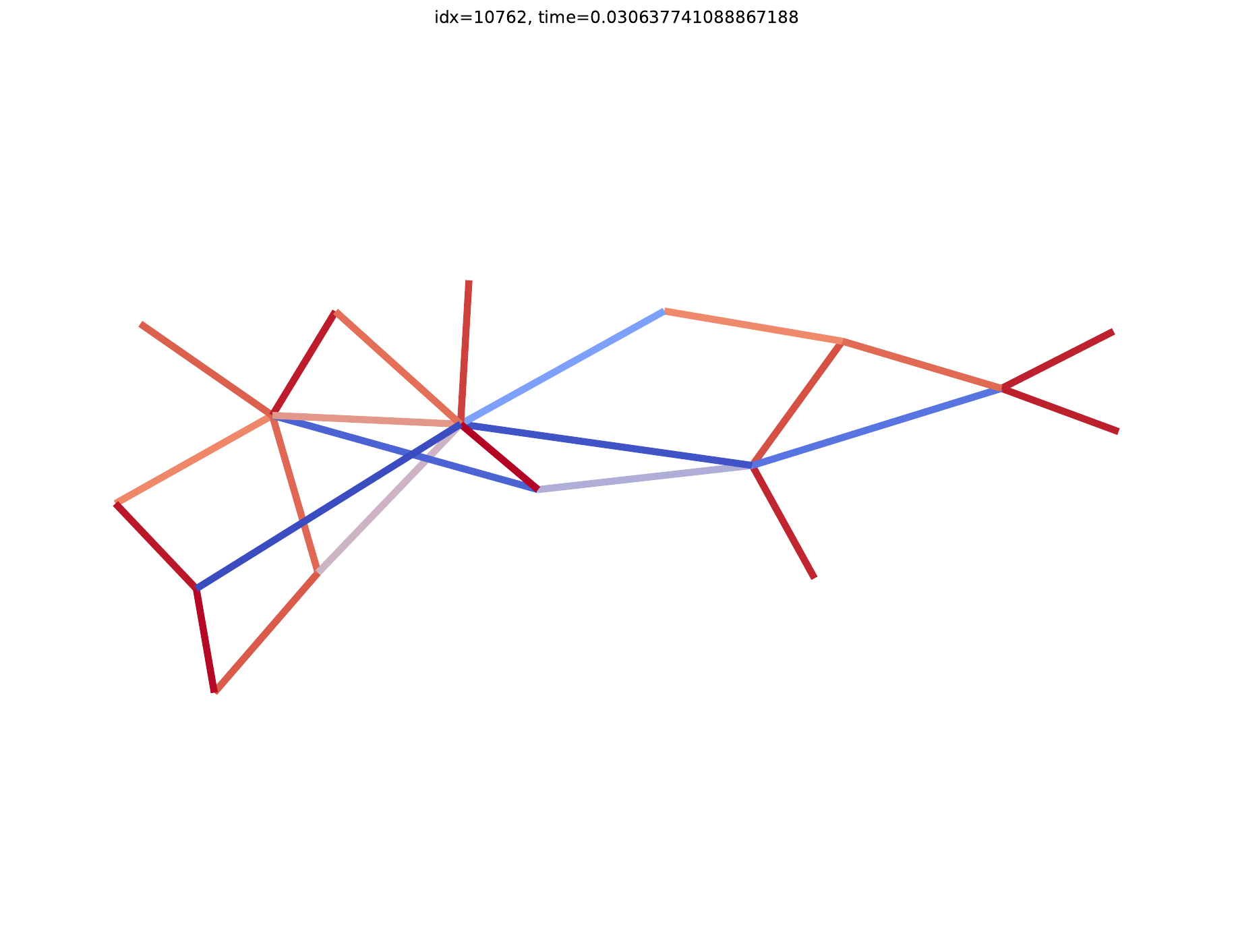} &
\imgcell{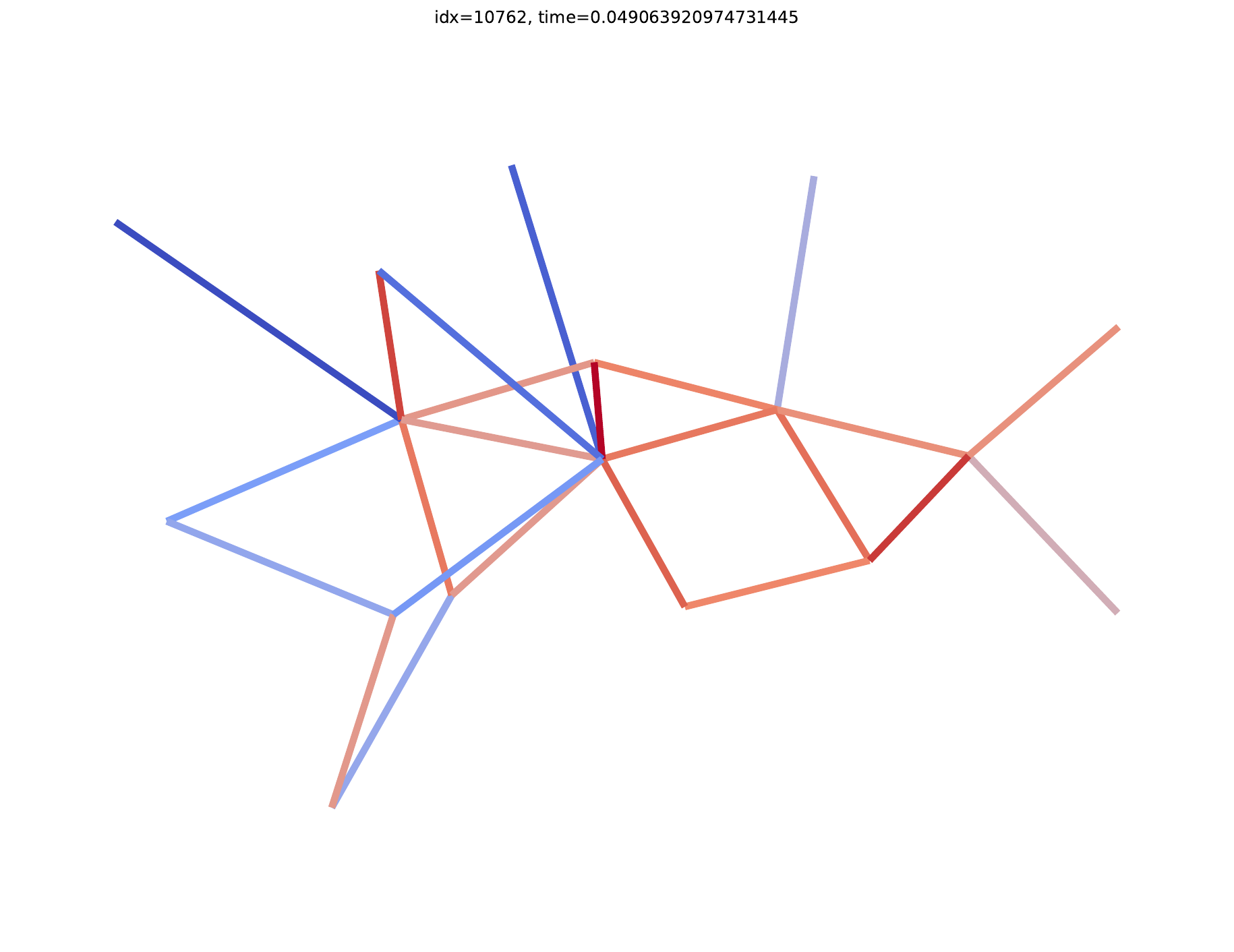} &
\imgcell{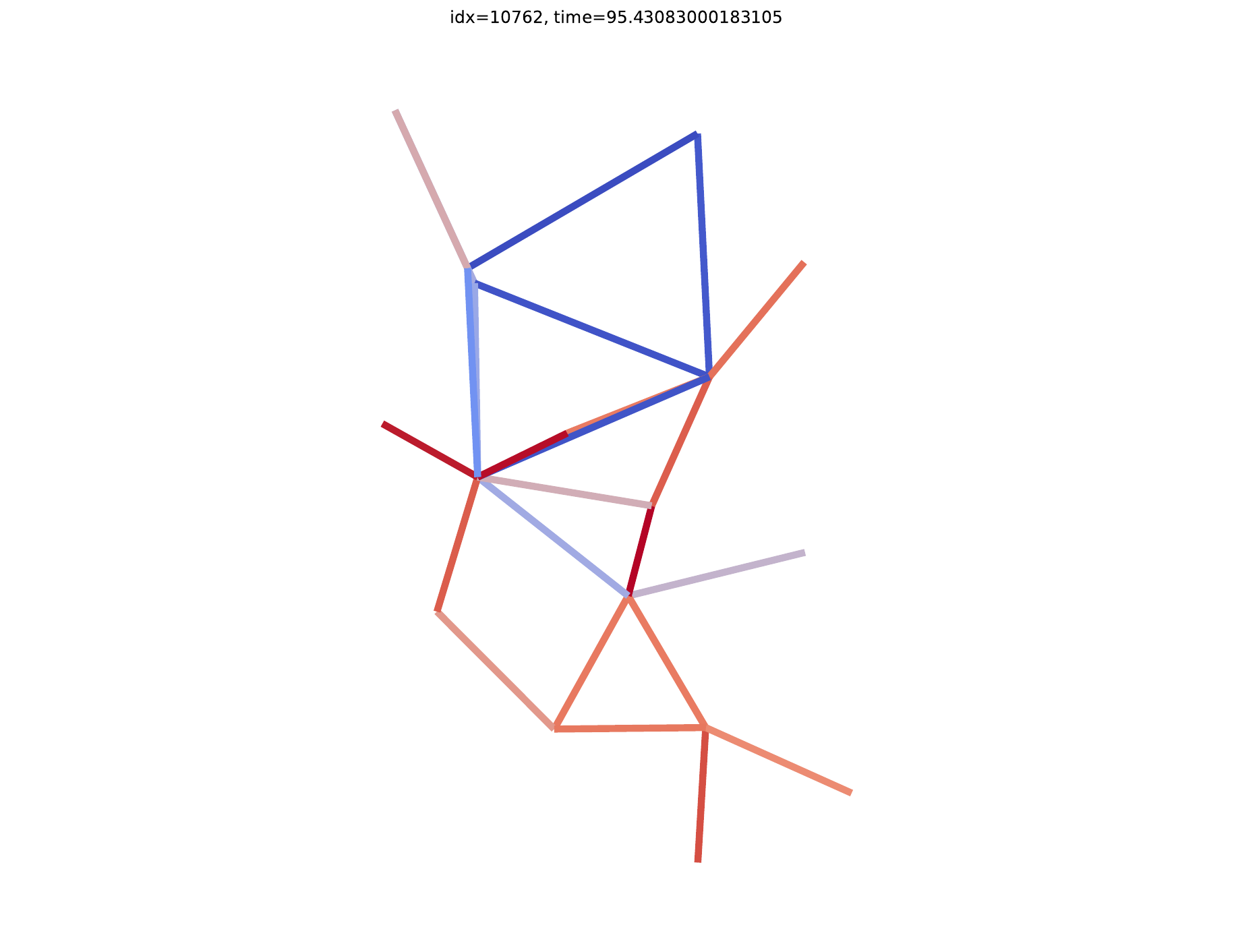} &
\imgcell{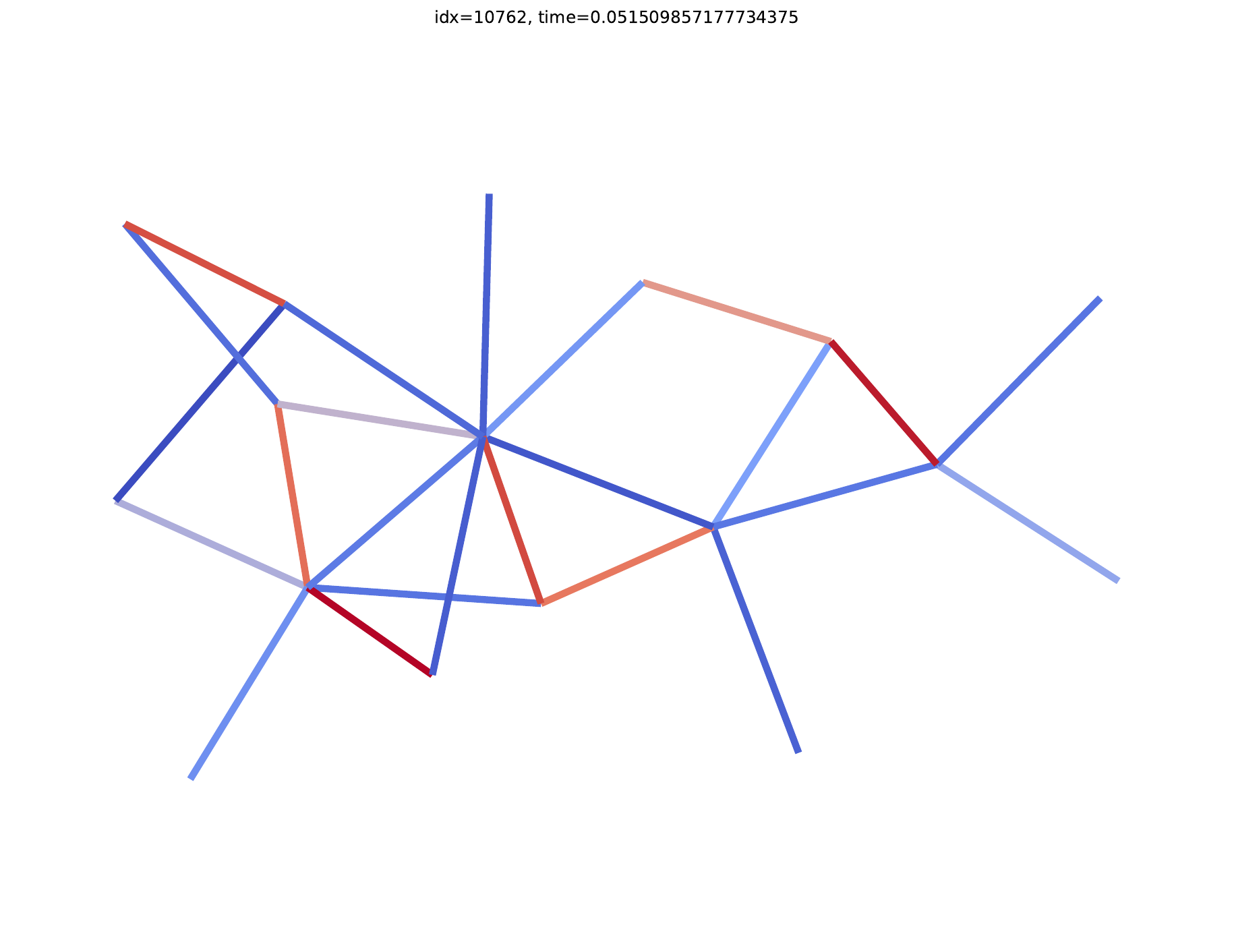} &
\imgcell{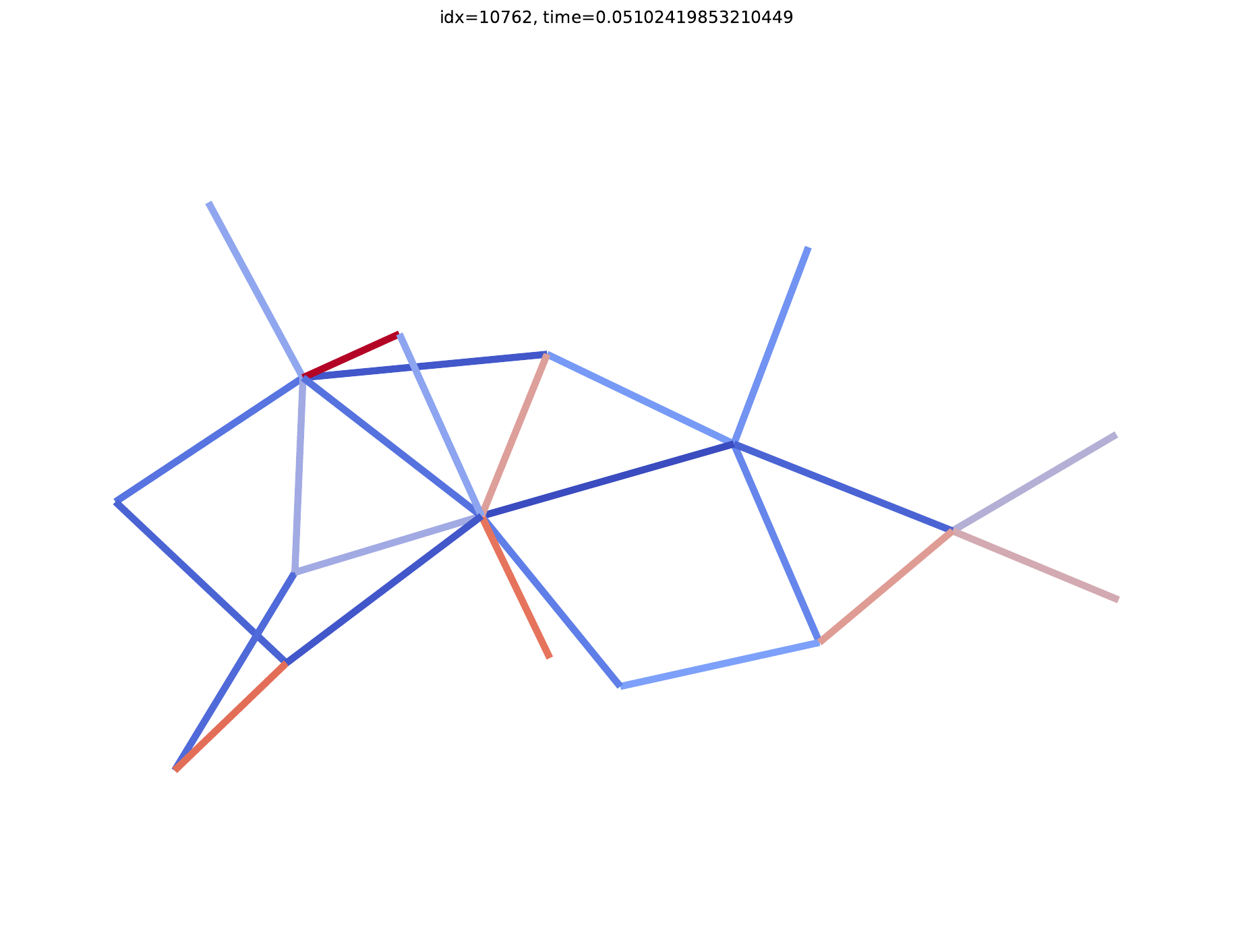} &
\imgcell{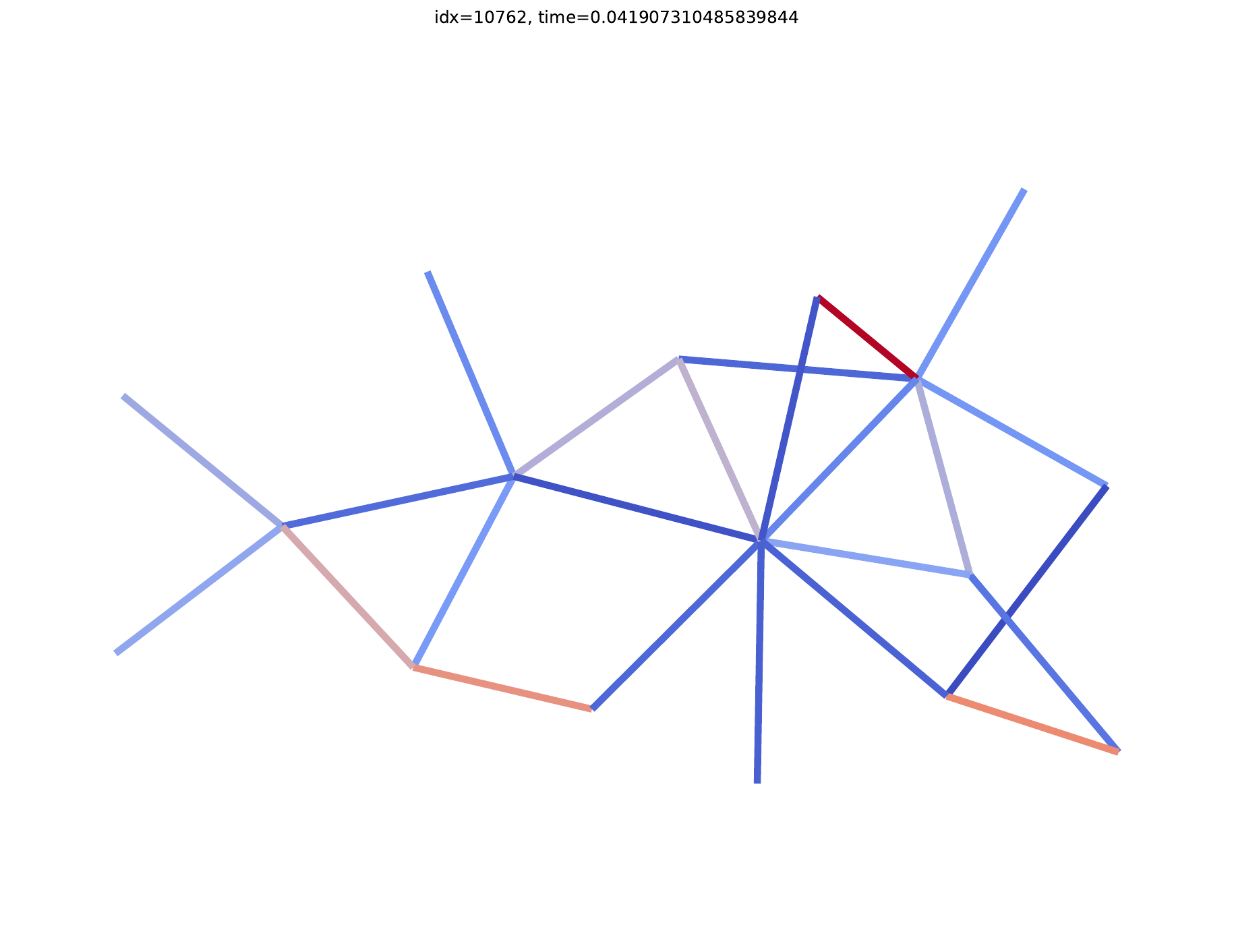} &
\imgcell{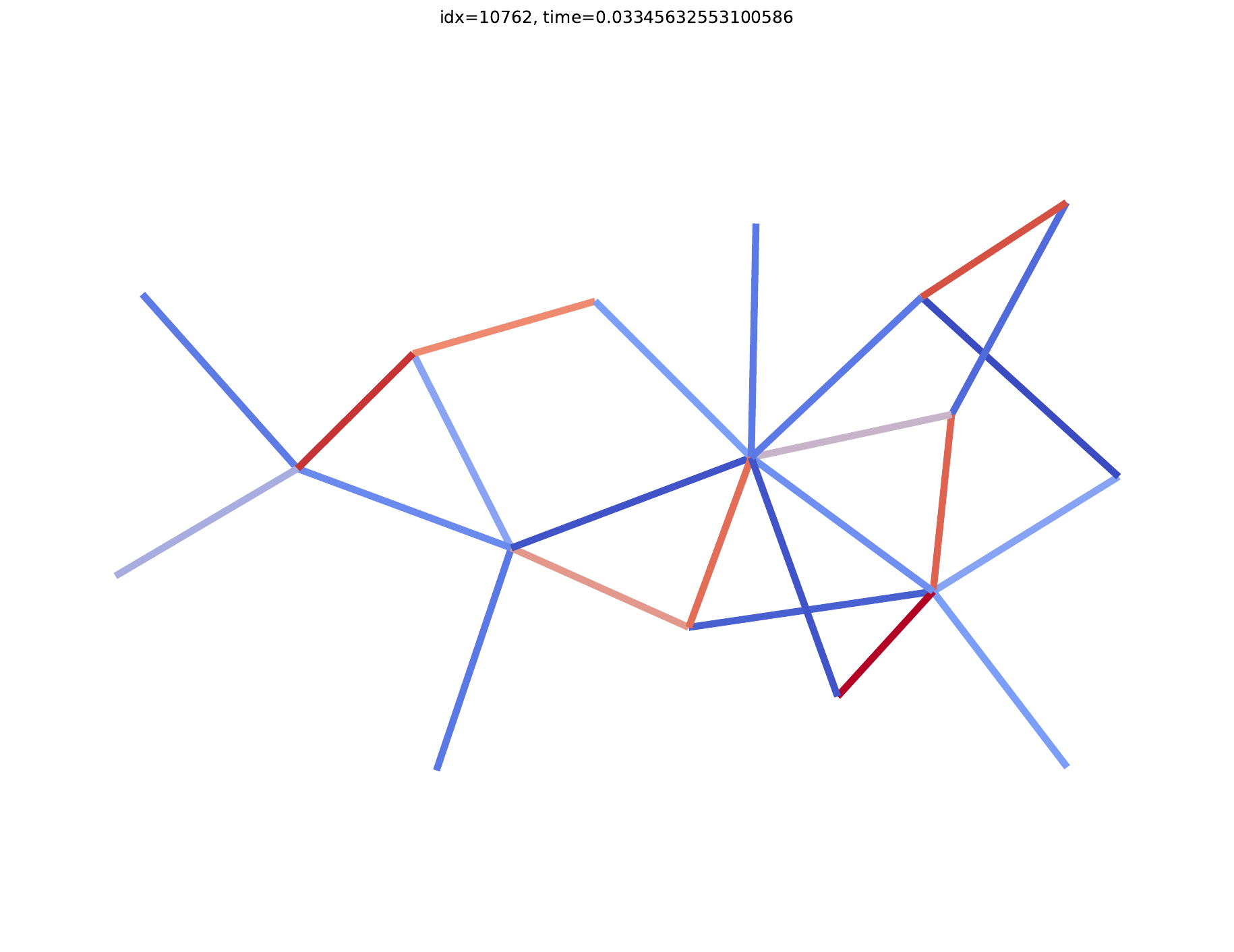} &
\imgcell{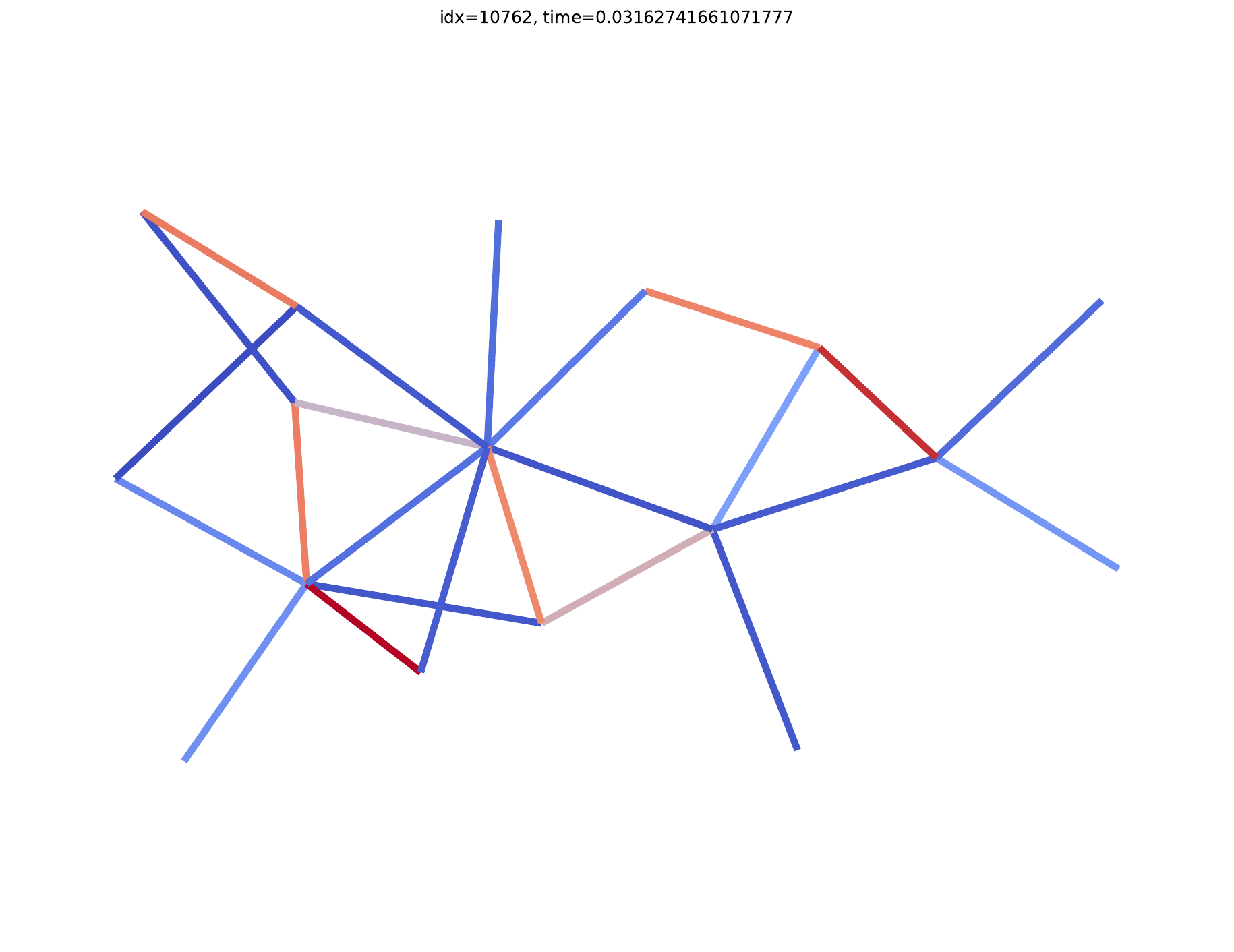} &
\imgcell{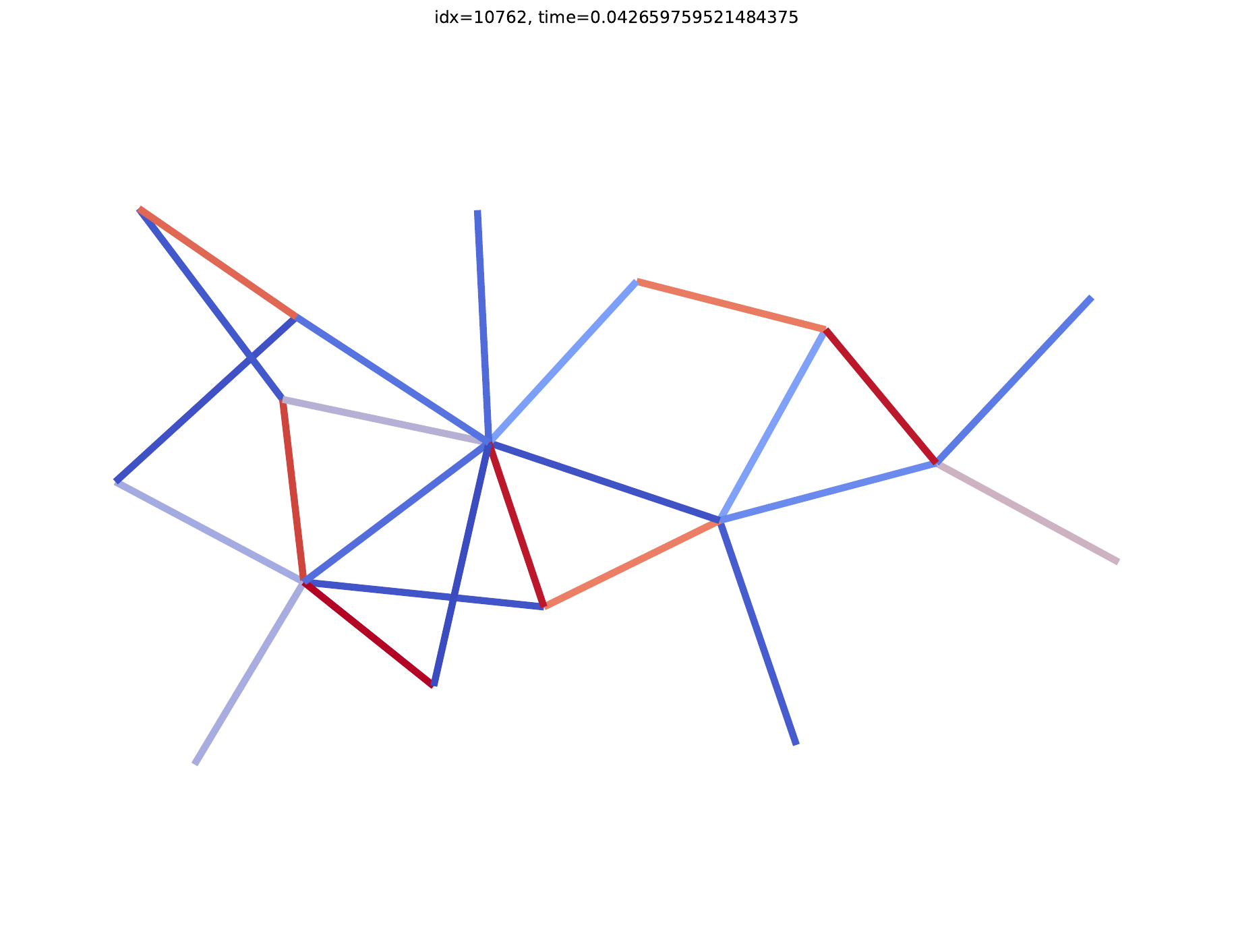} &
\imgcell{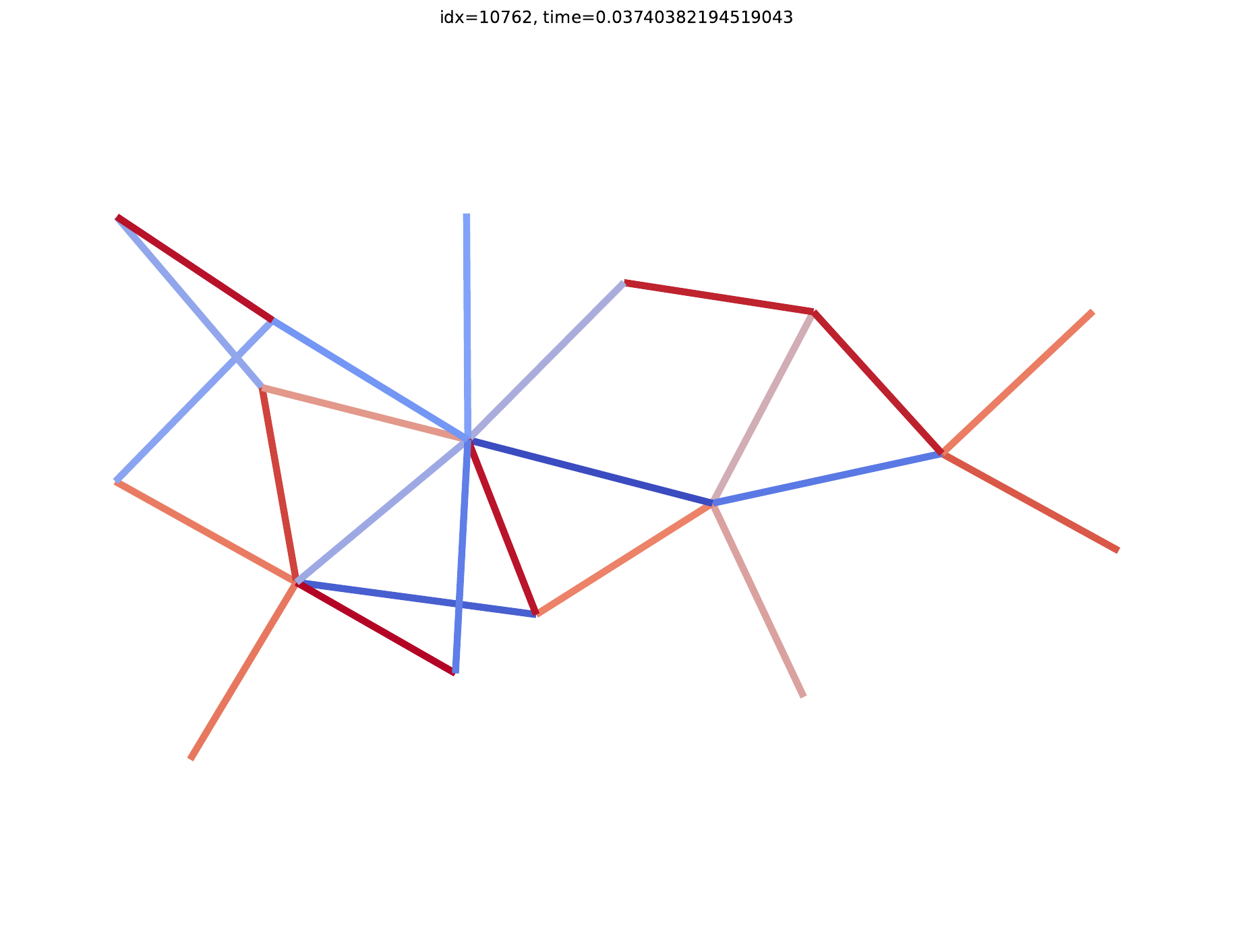} \\

&
t = 0.00s &
t = 0.22s &
t = 0.03s &
t = 0.05s &
t = 95.43s &
t = 0.05s &
t = 0.05s &
t = 0.04s &
t = 0.03s &
t = 0.03s &
t = 0.04s &
t = 0.04s \\

\makecell{\bfseries grafo731.14\\N = 60\\M = 87} &
\imgcell{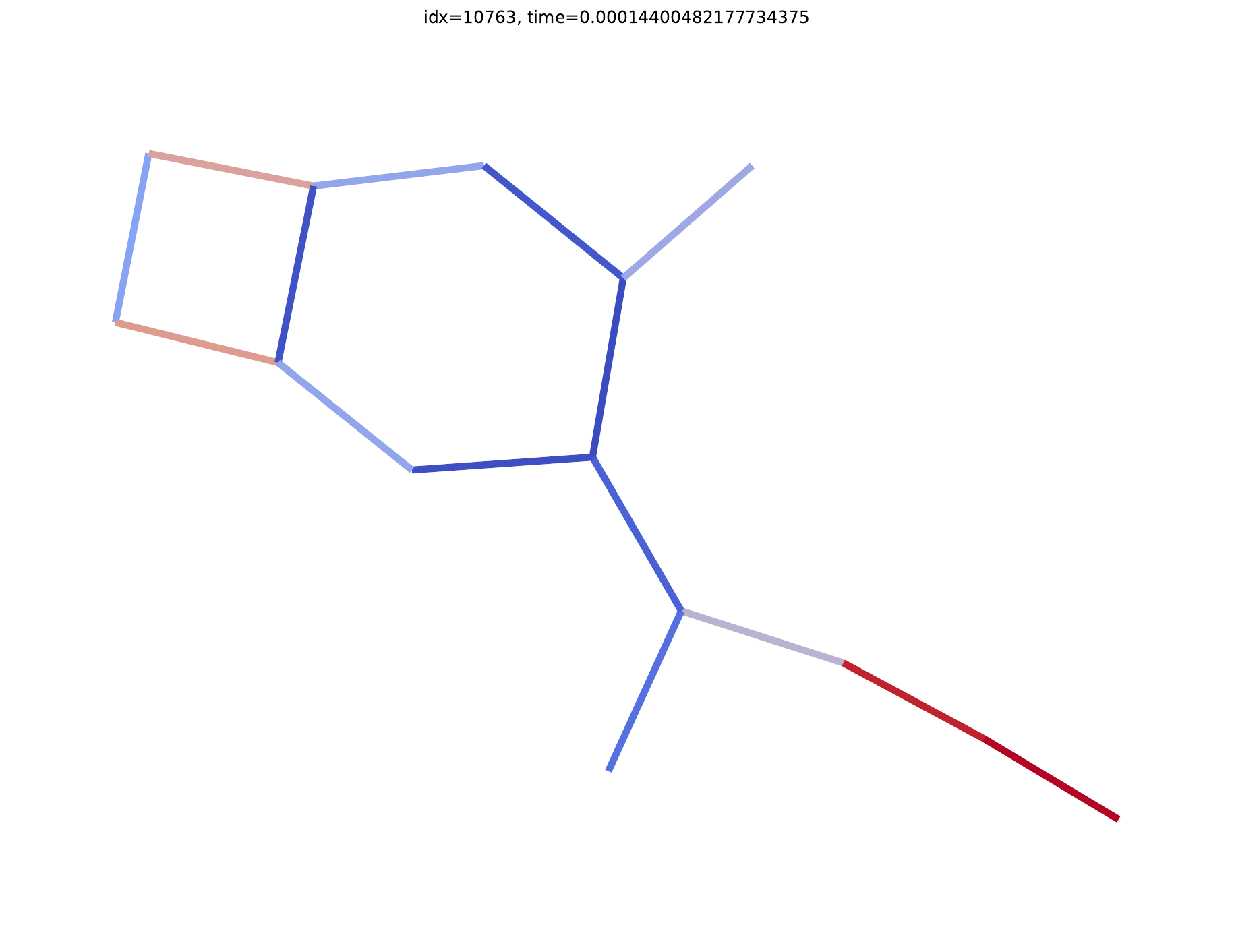} &
\imgcell{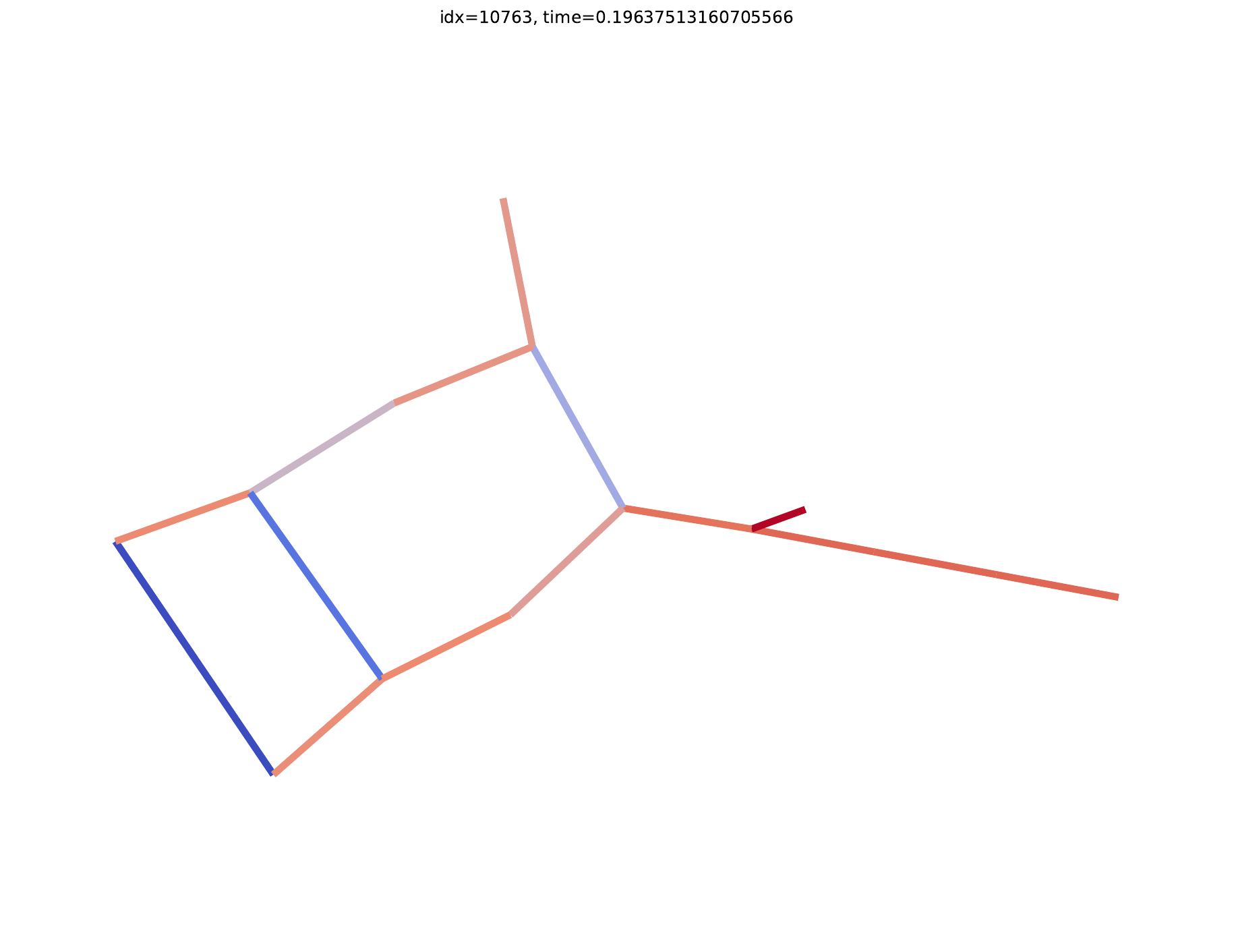} &
\imgcell{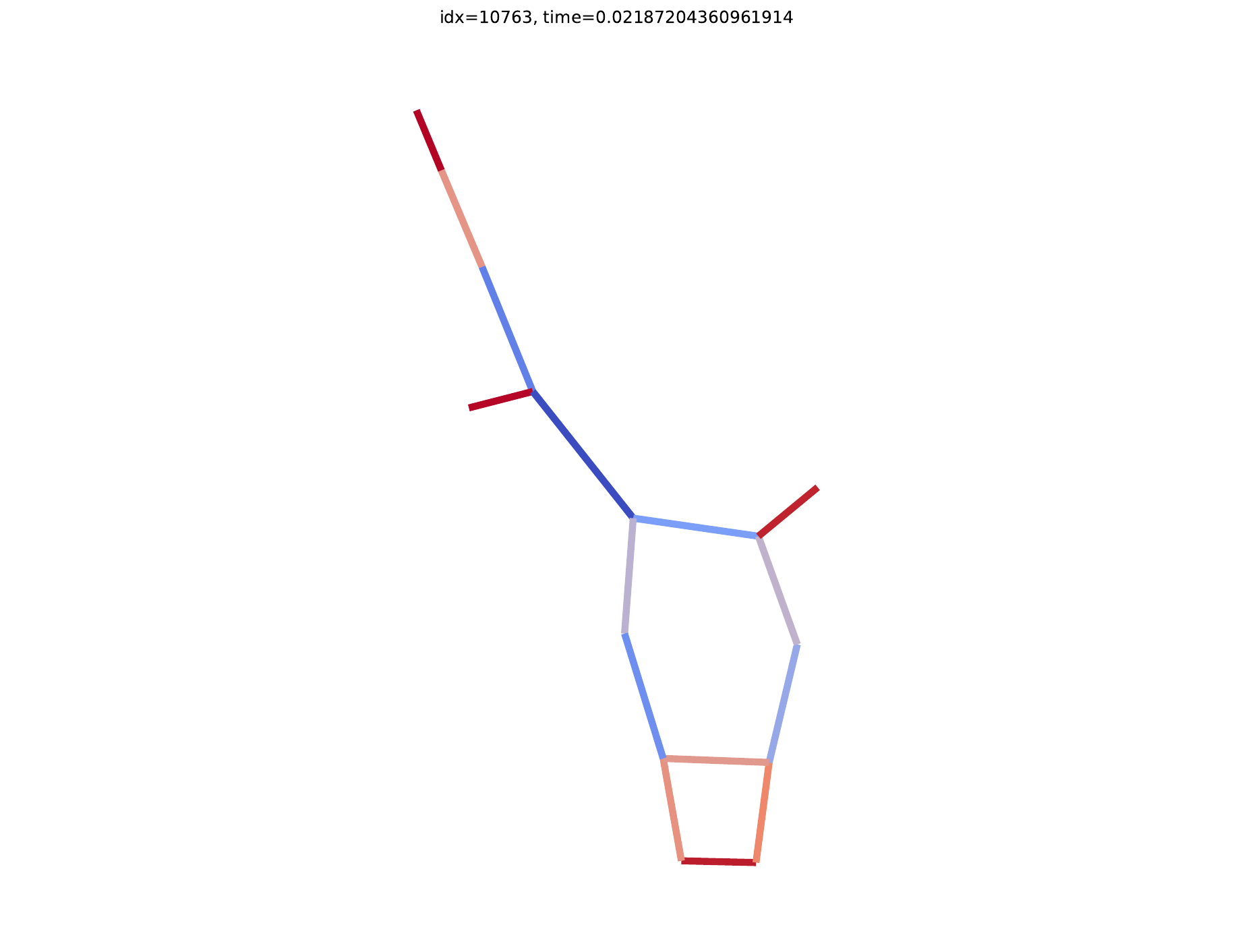} &
\imgcell{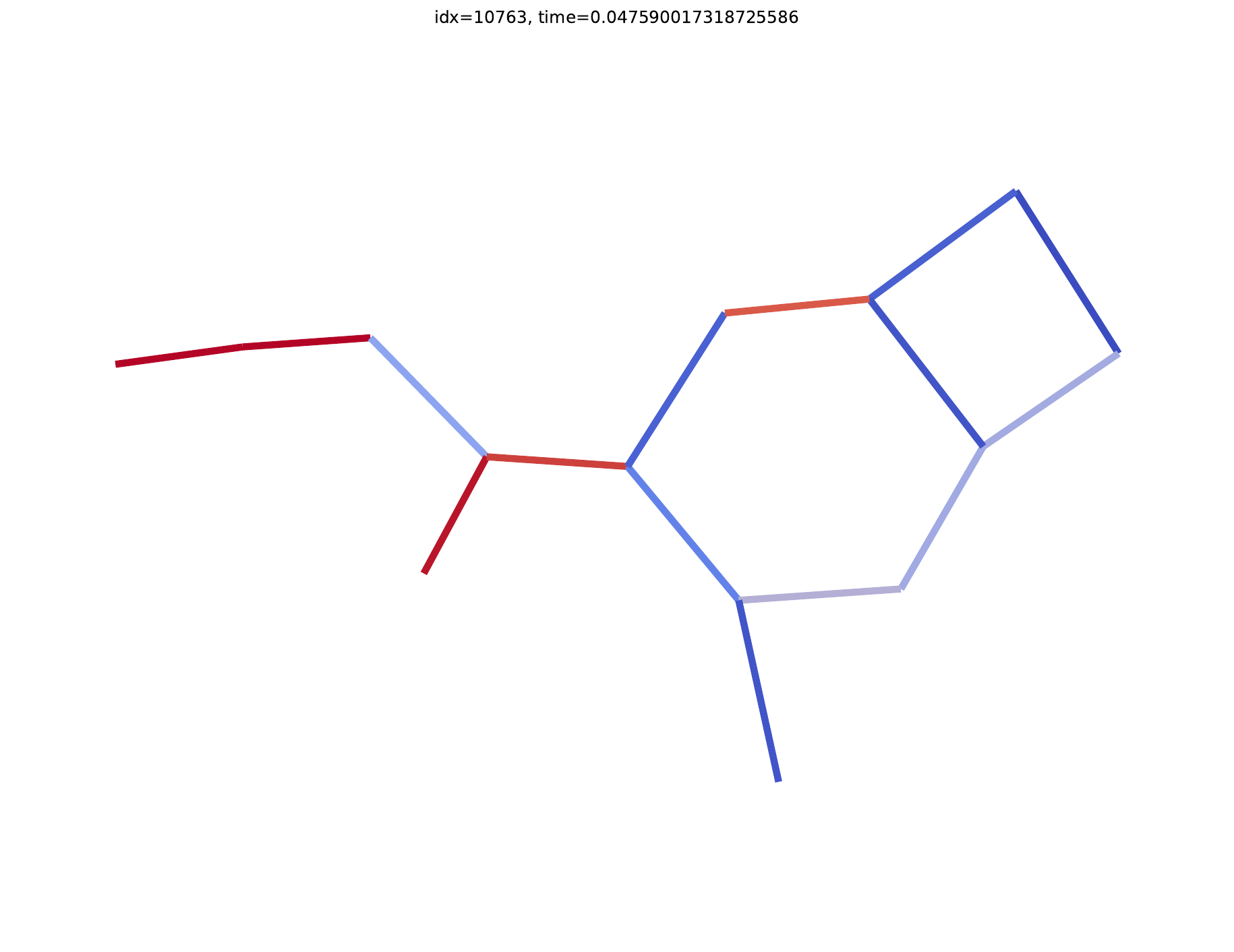} &
\imgcell{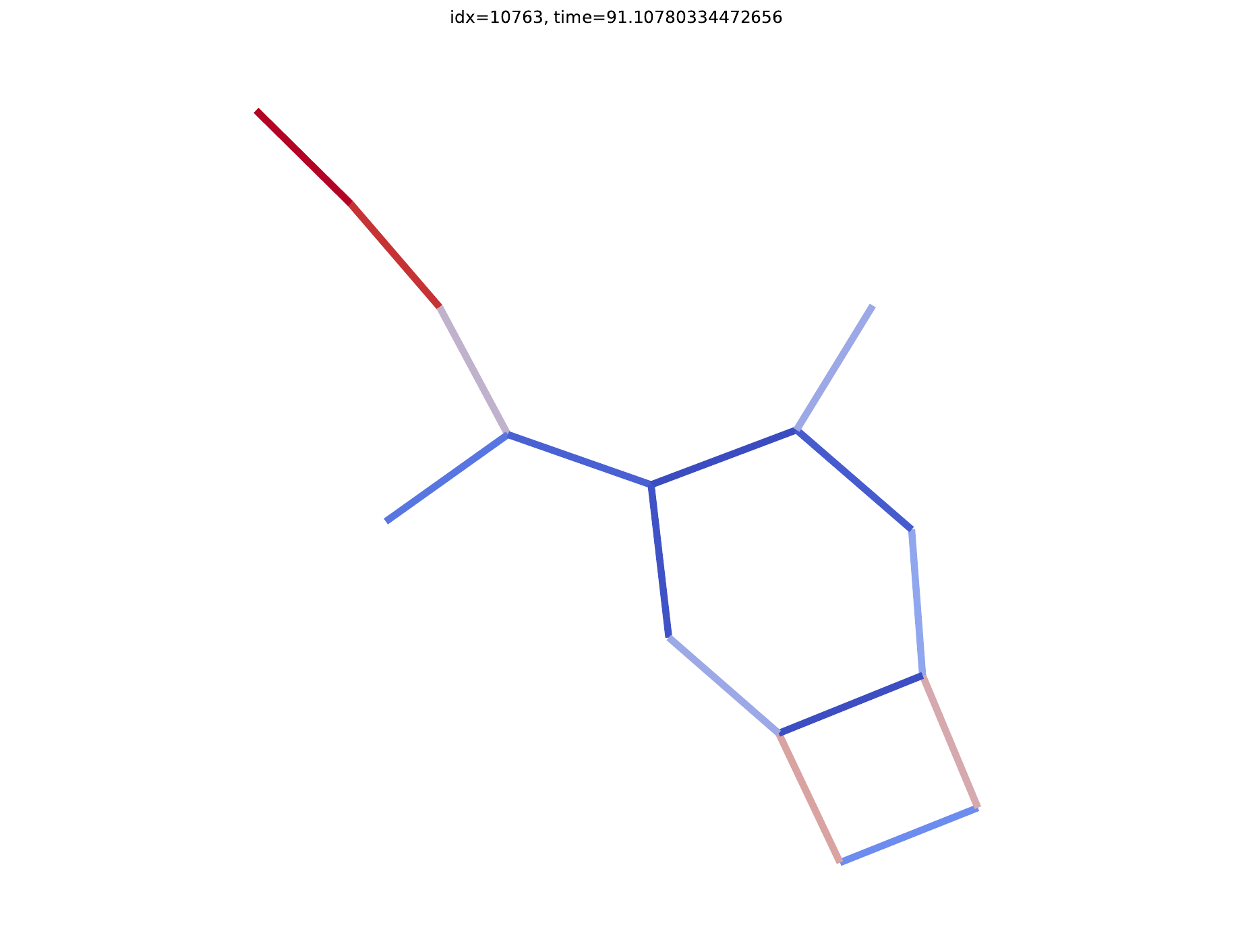} &
\imgcell{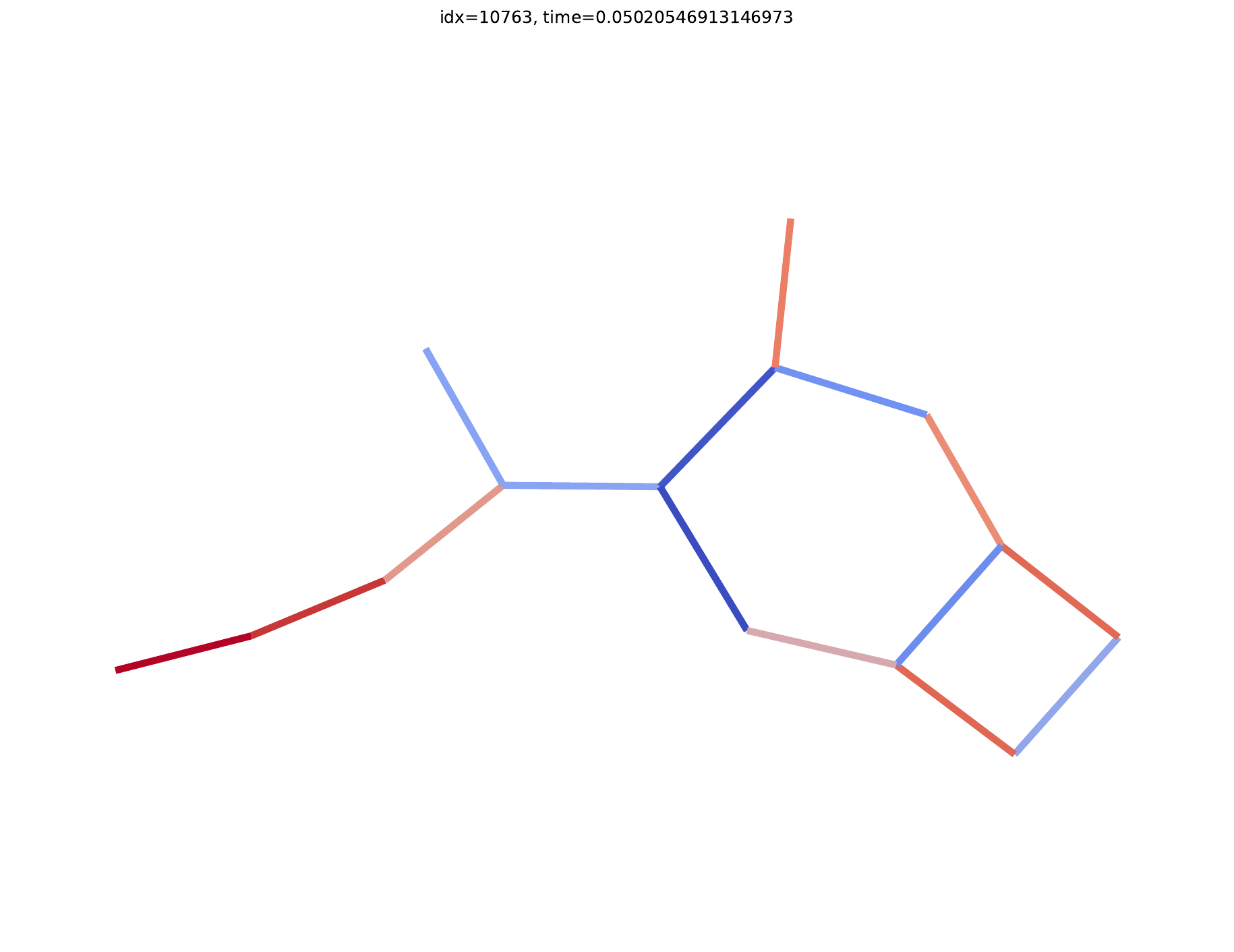} &
\imgcell{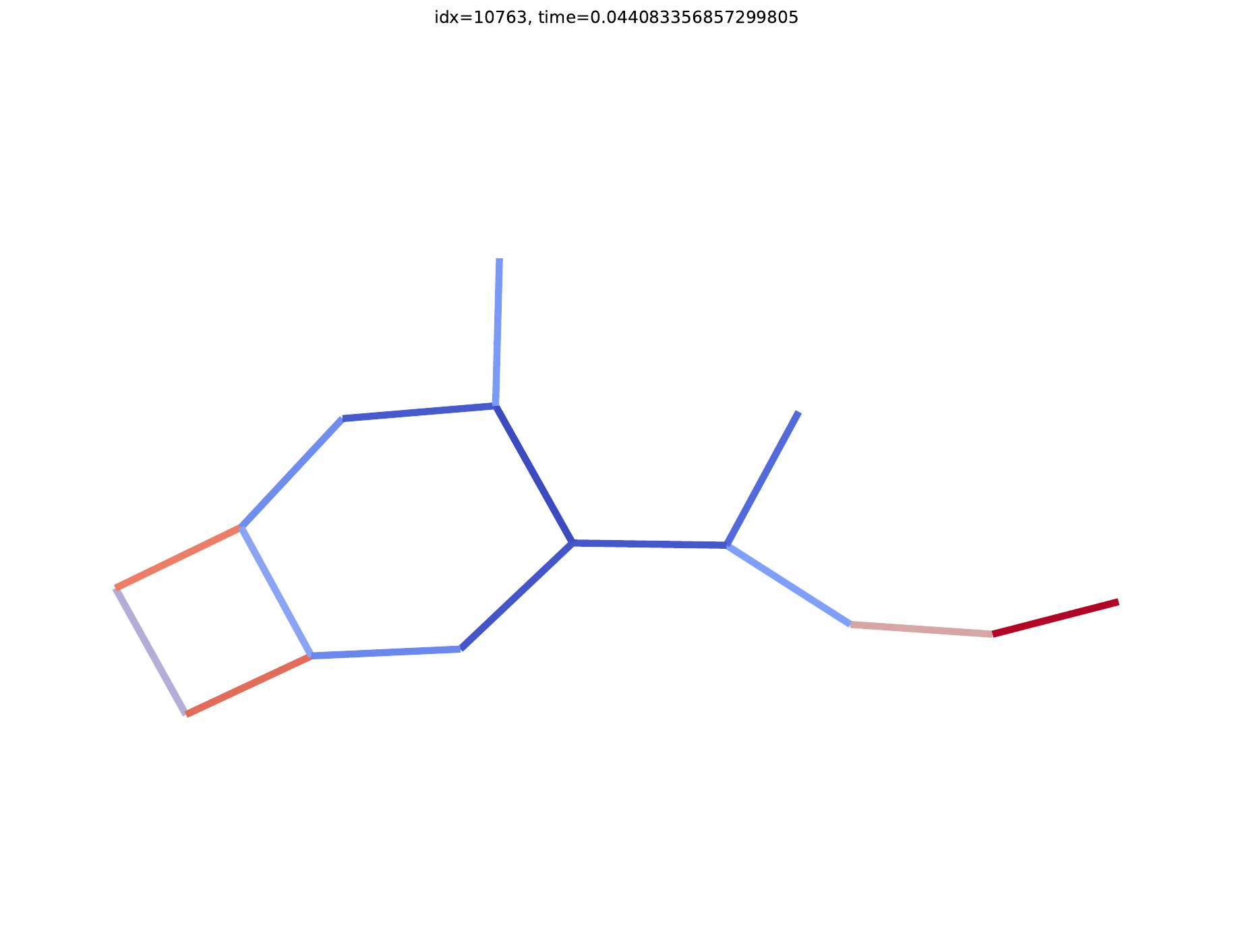} &
\imgcell{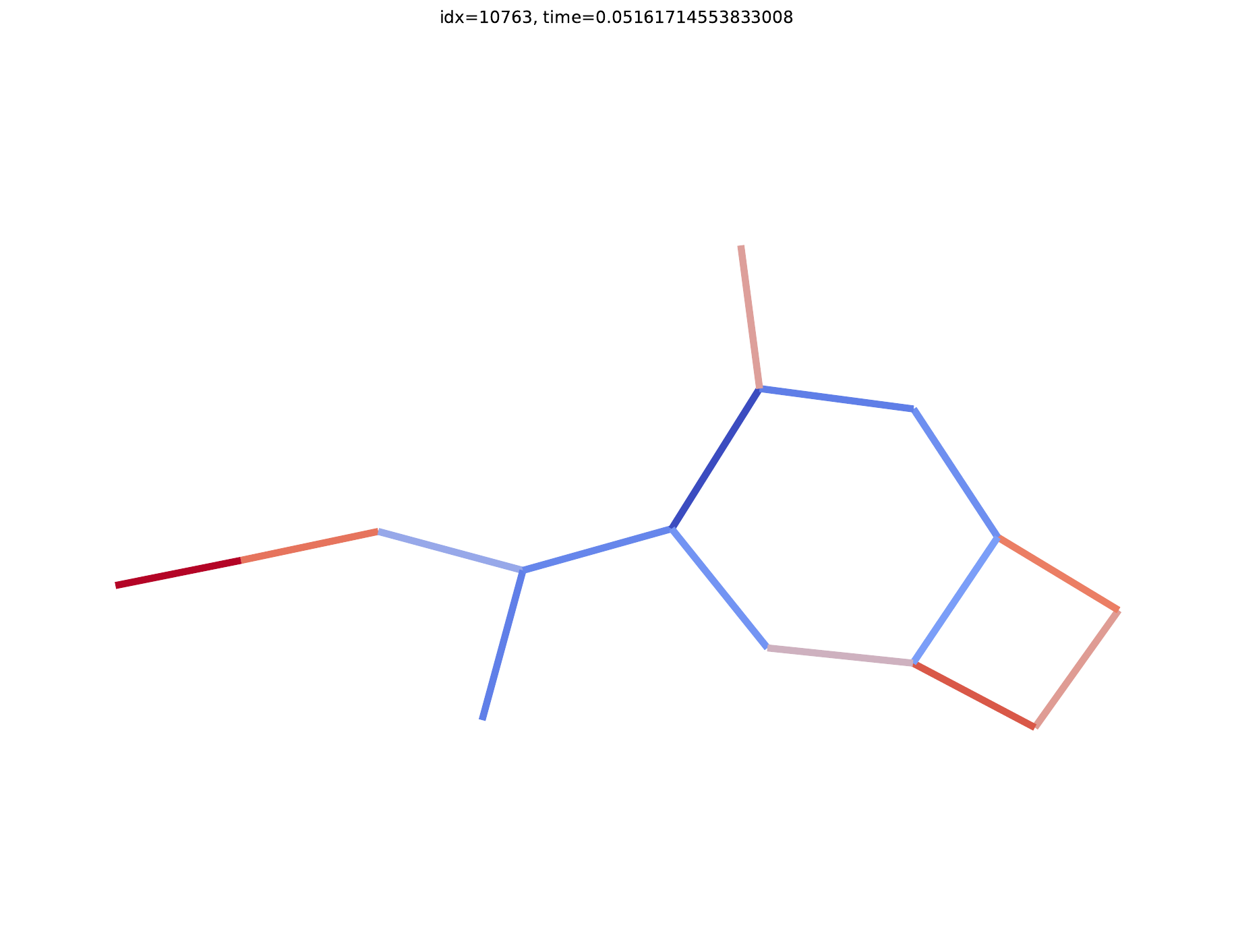} &
\imgcell{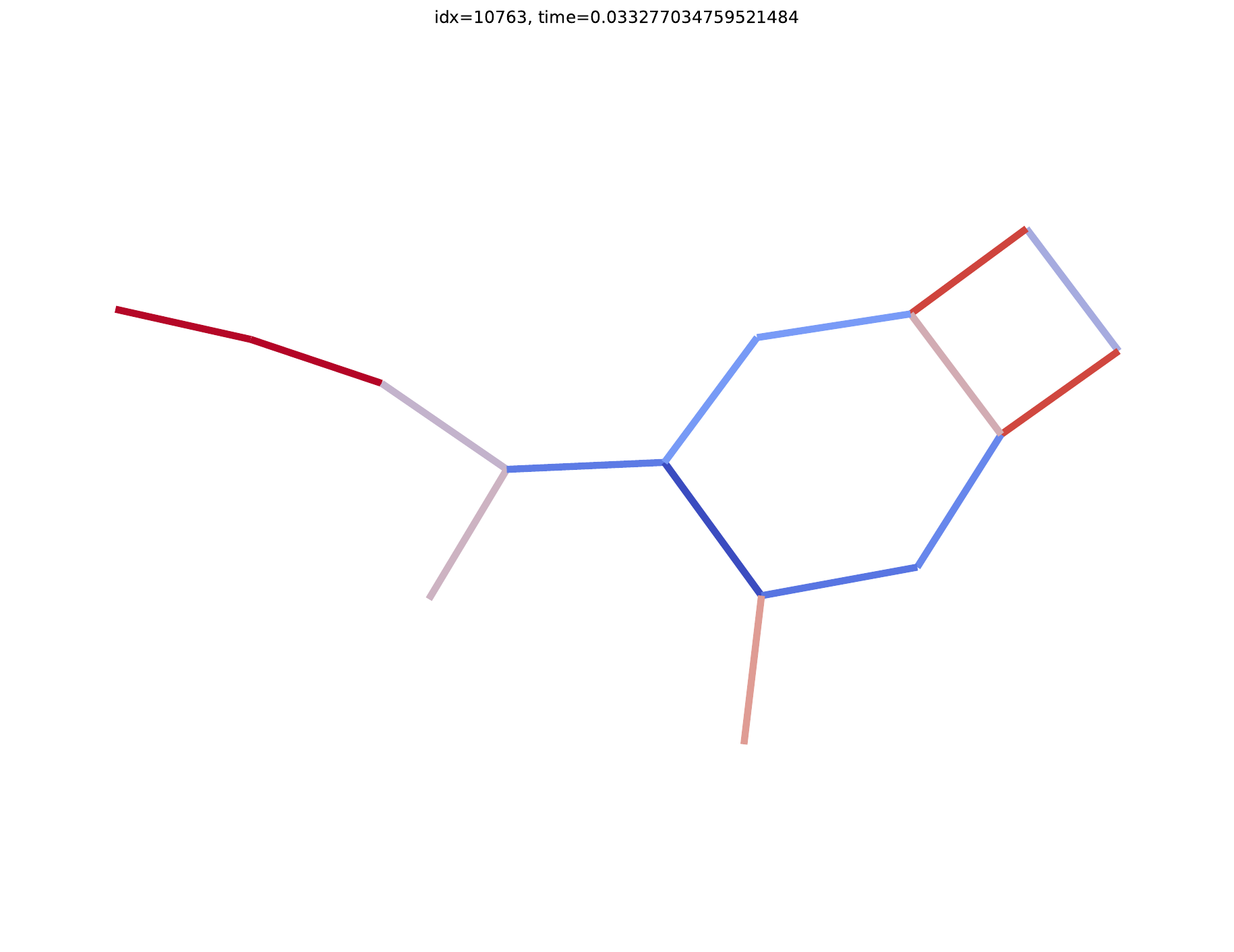} &
\imgcell{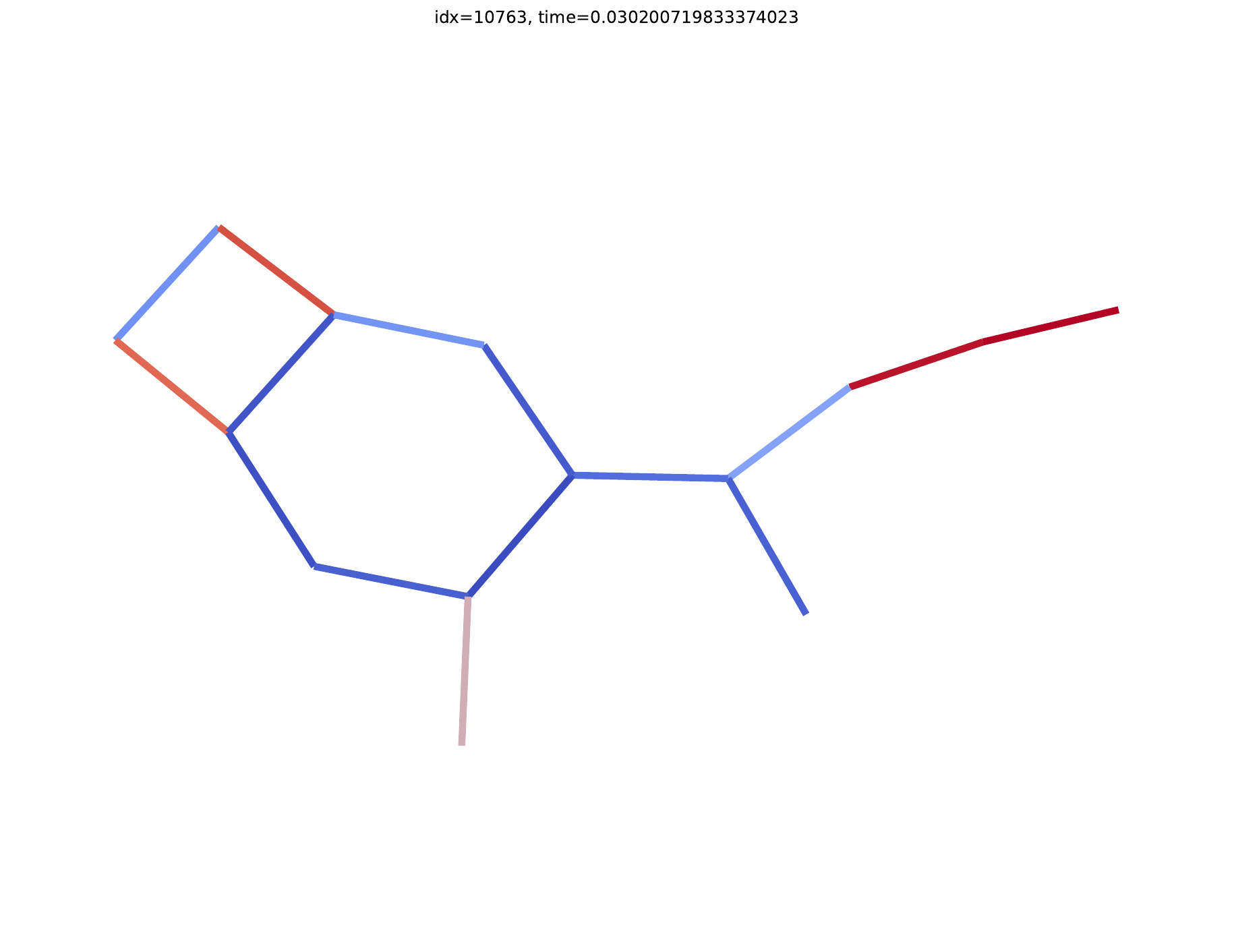} &
\imgcell{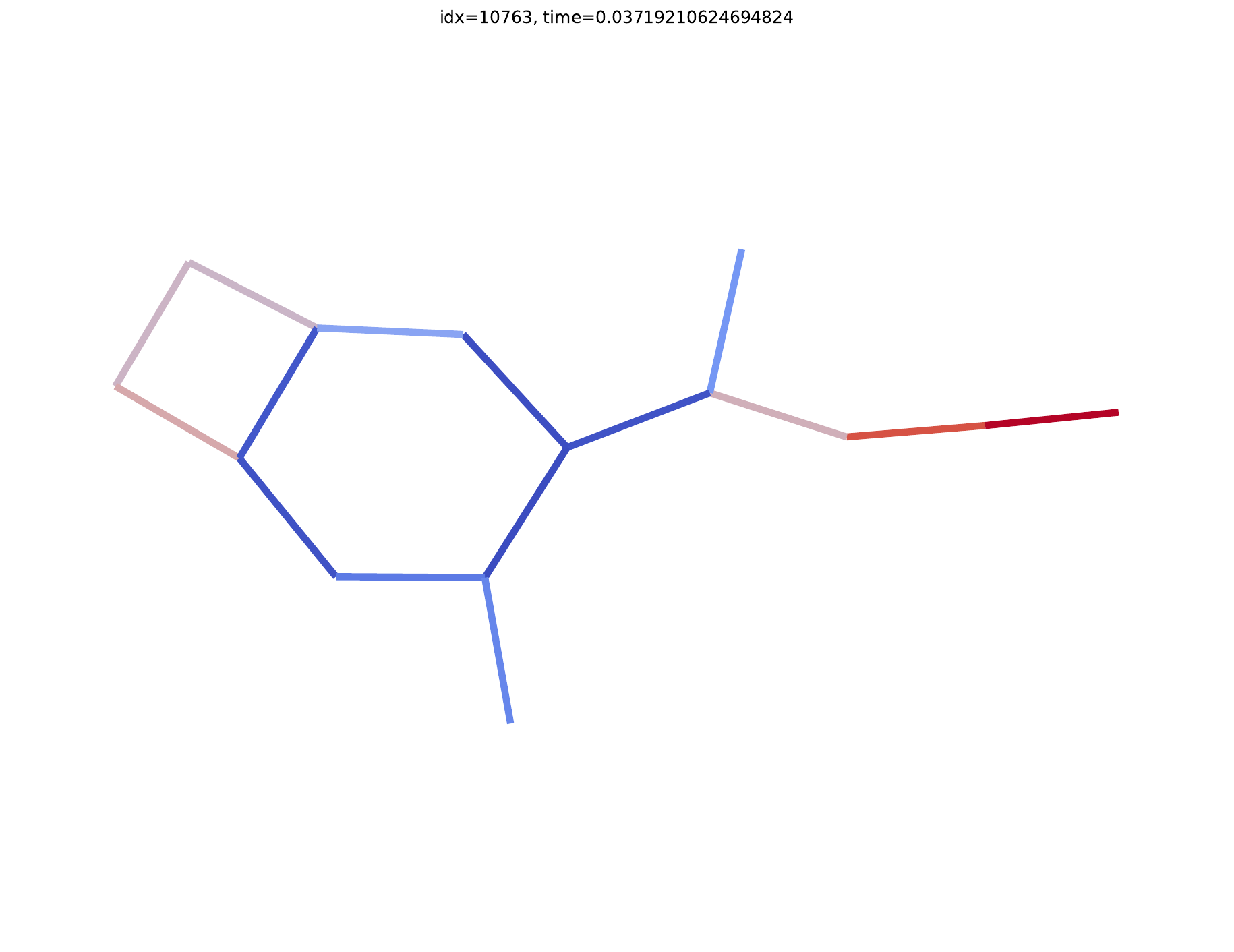} &
\imgcell{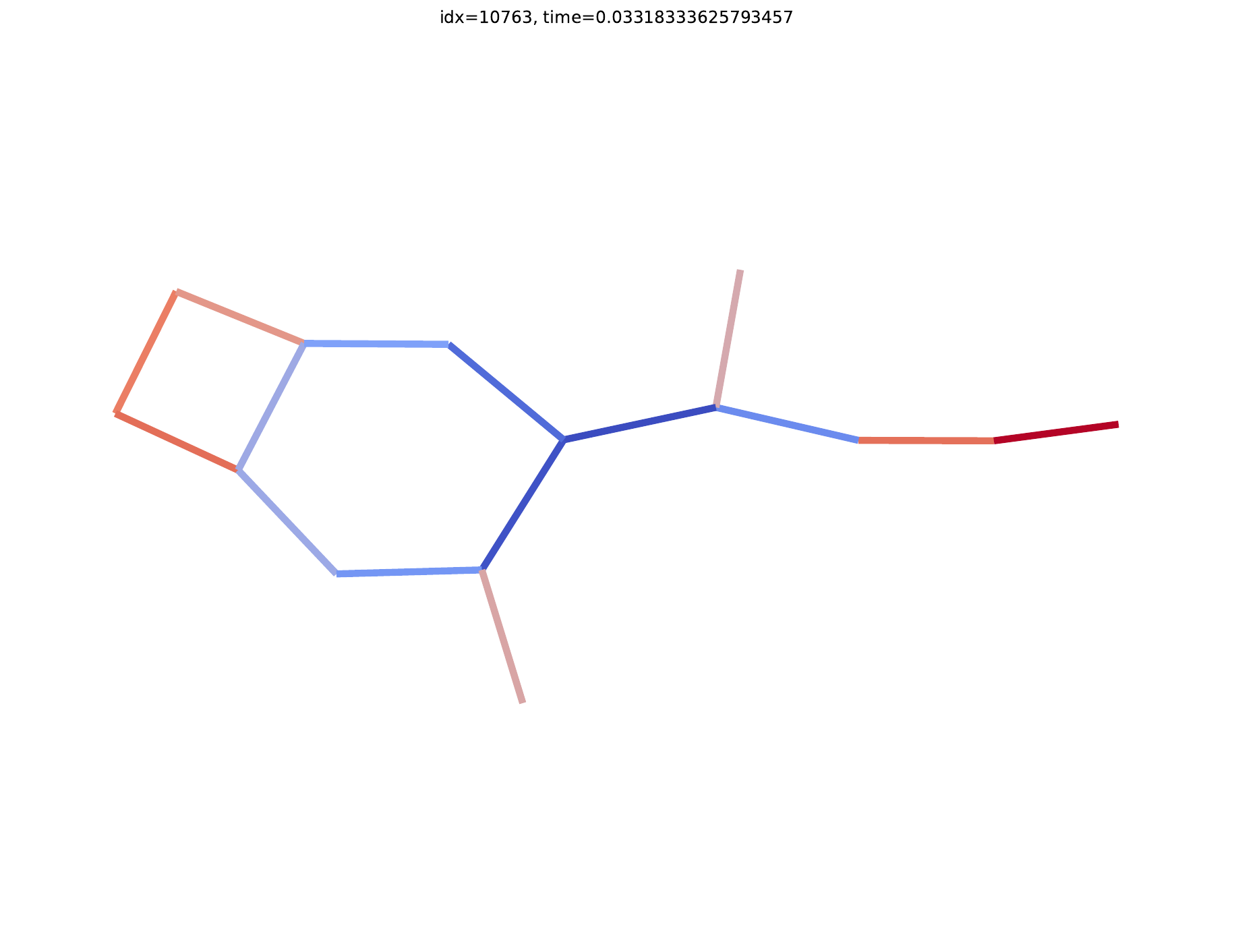} \\

&
t = 0.00s &
t = 0.20s &
t = 0.02s &
t = 0.05s &
t = 91.11s &
t = 0.05s &
t = 0.04s &
t = 0.04s &
t = 0.03s &
t = 0.03s &
t = 0.04s &
t = 0.03s \\

\makecell{\bfseries grafo1044.23\\N = 100\\M = 124} &
\imgcell{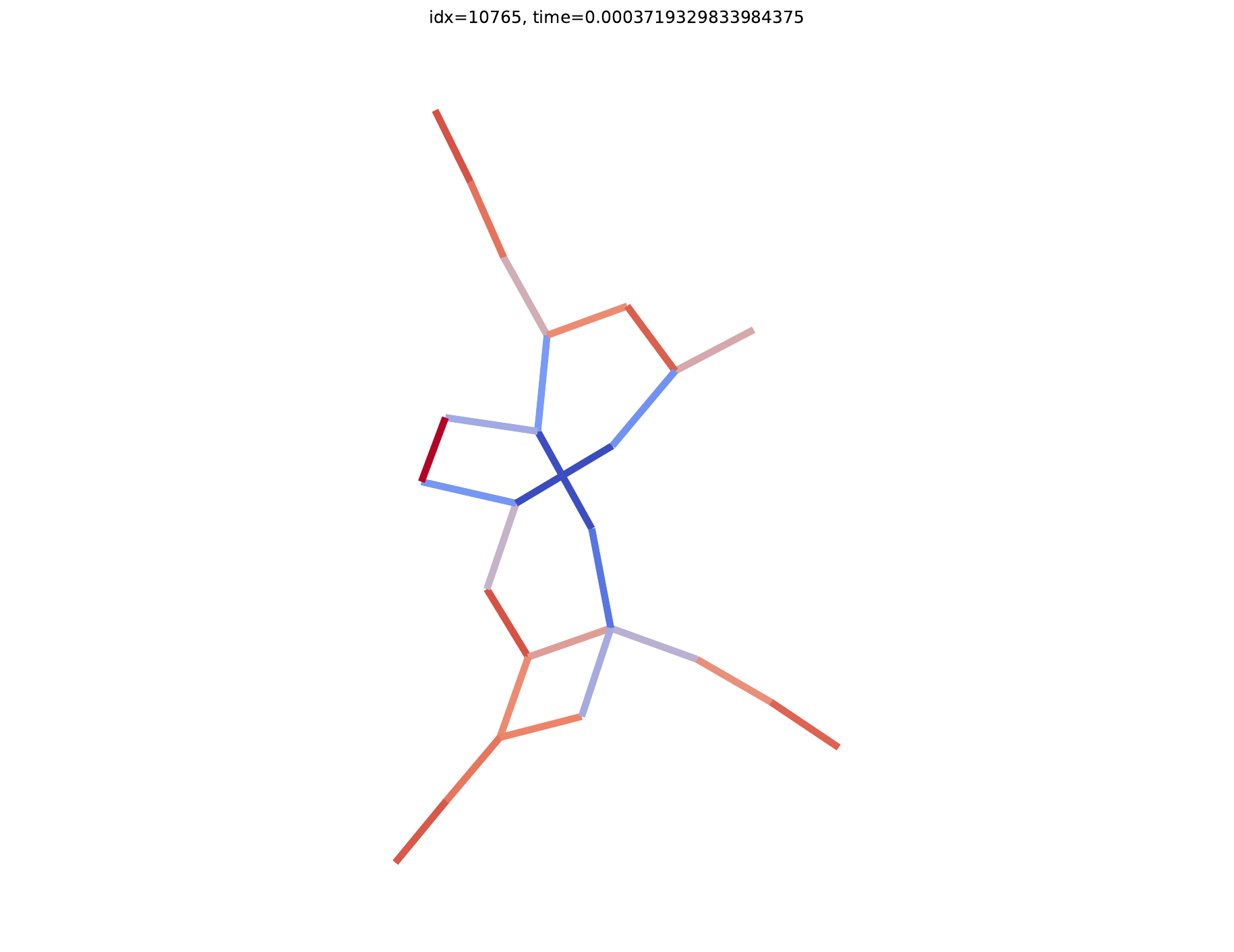} &
\imgcell{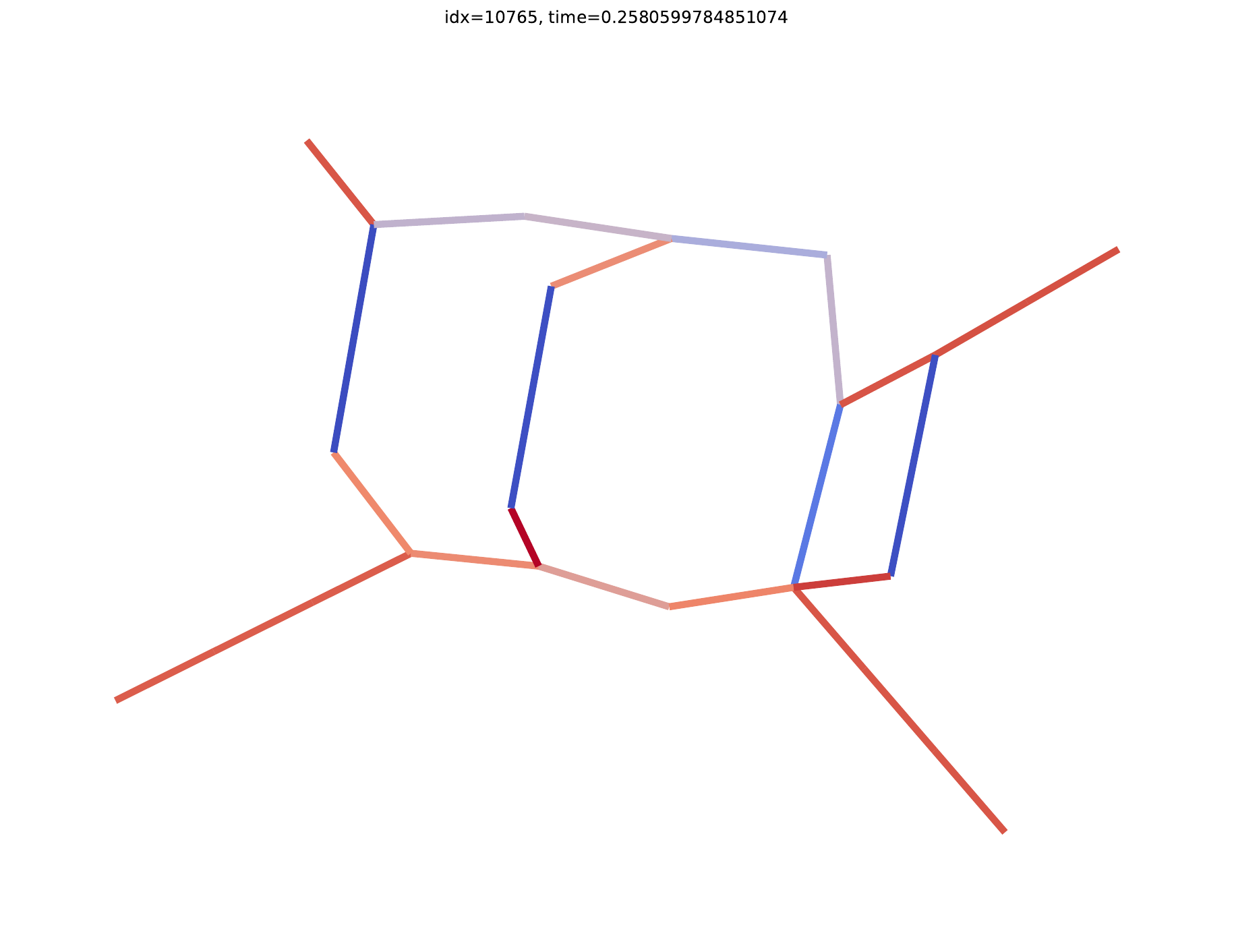} &
\imgcell{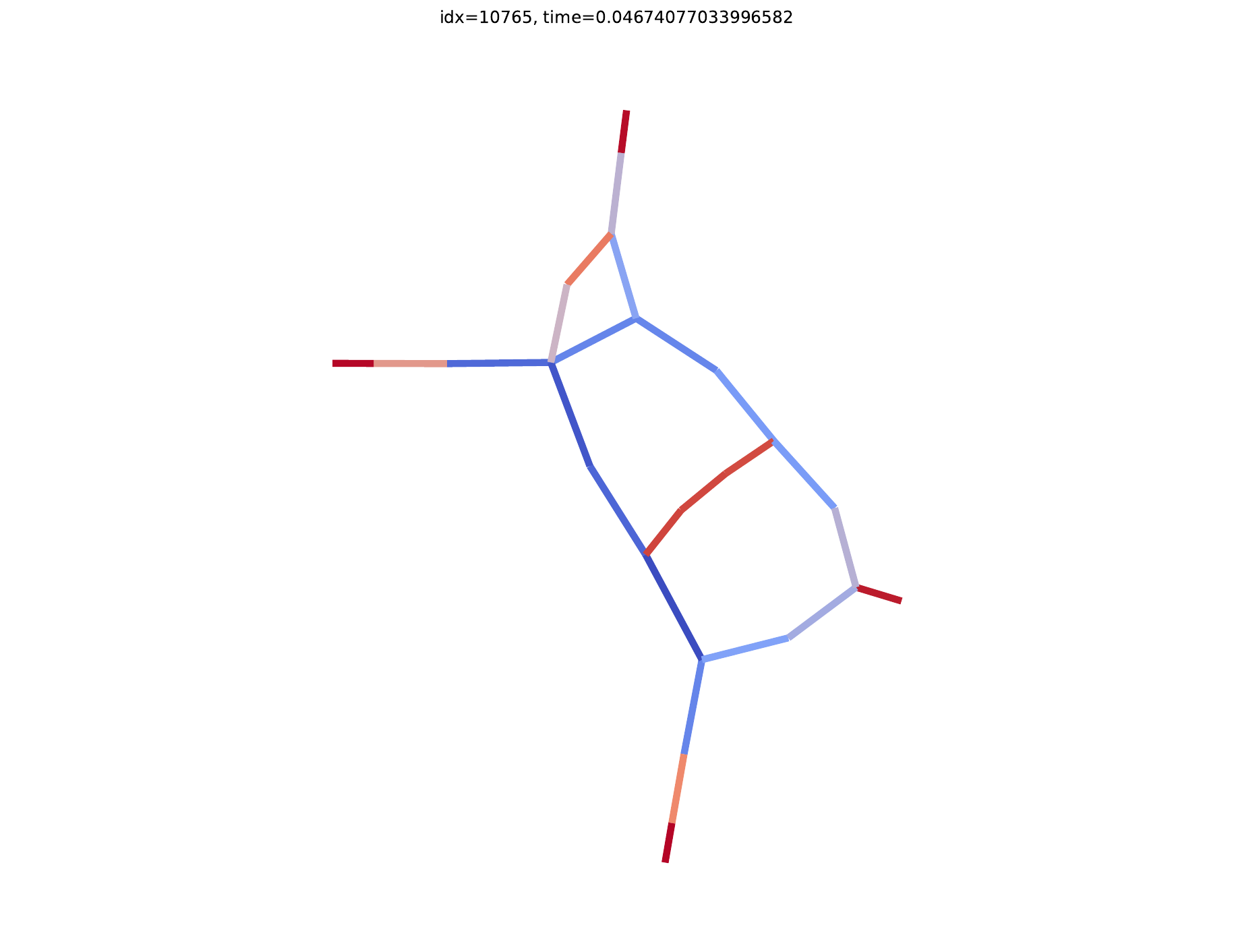} &
\imgcell{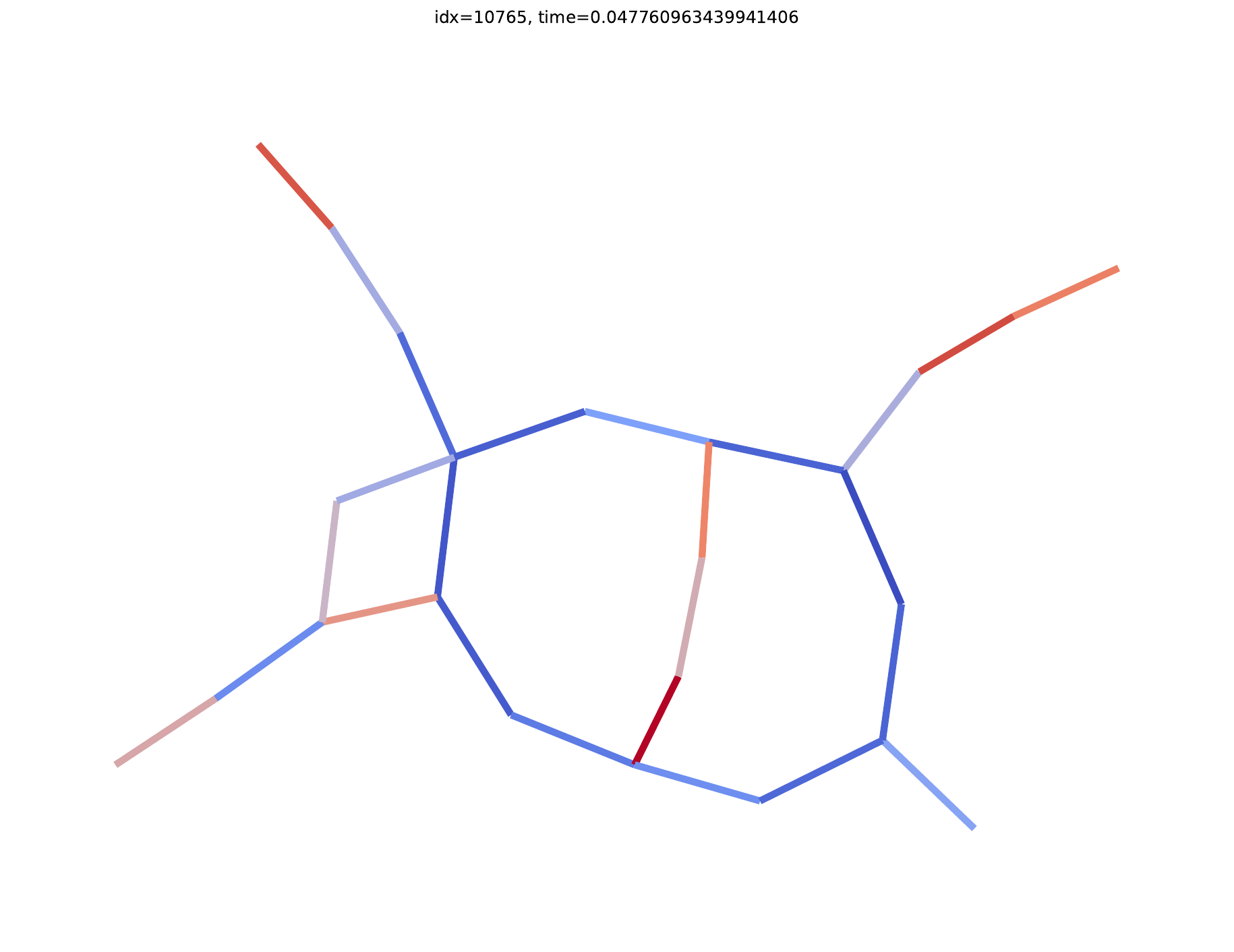} &
\imgcell{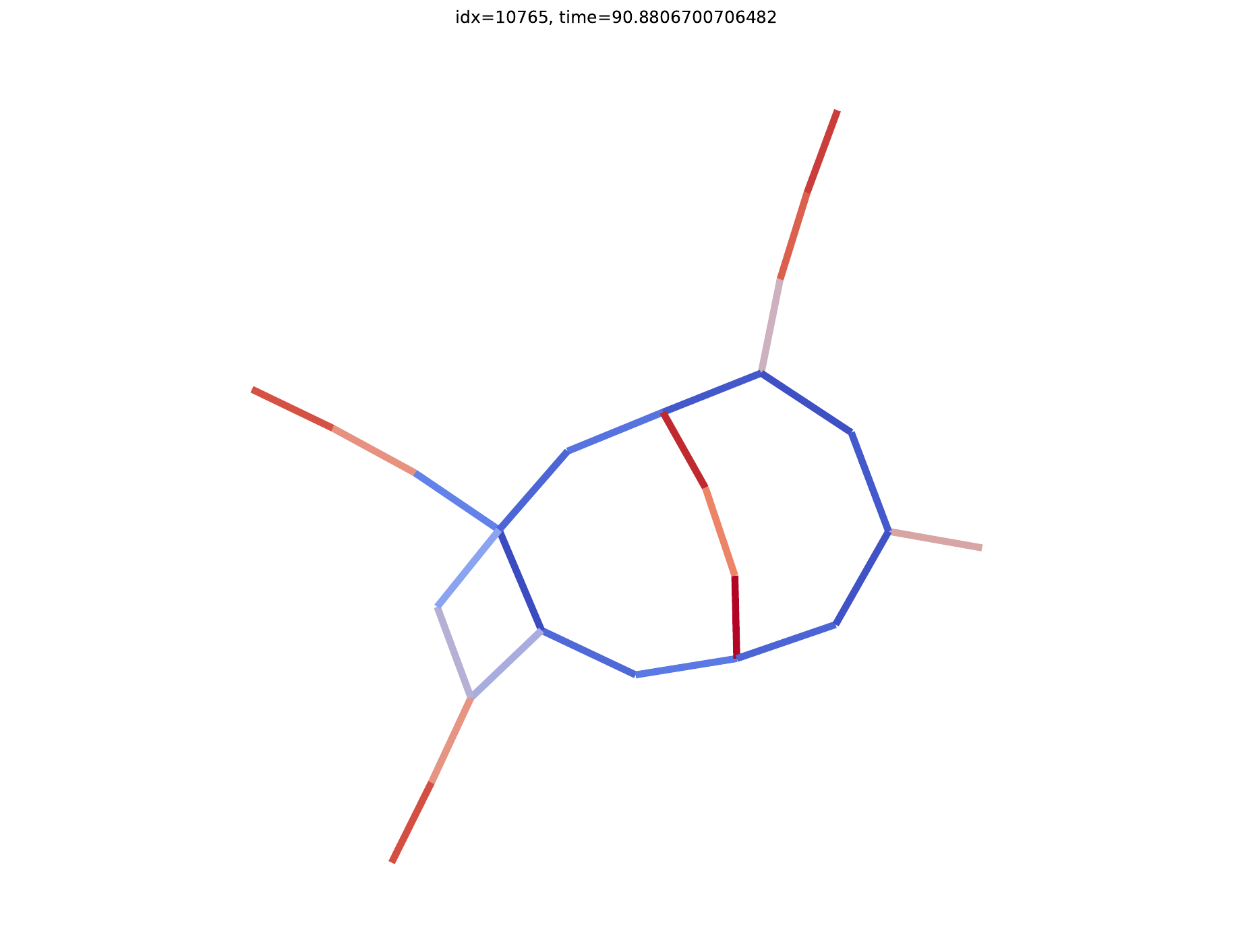} &
\imgcell{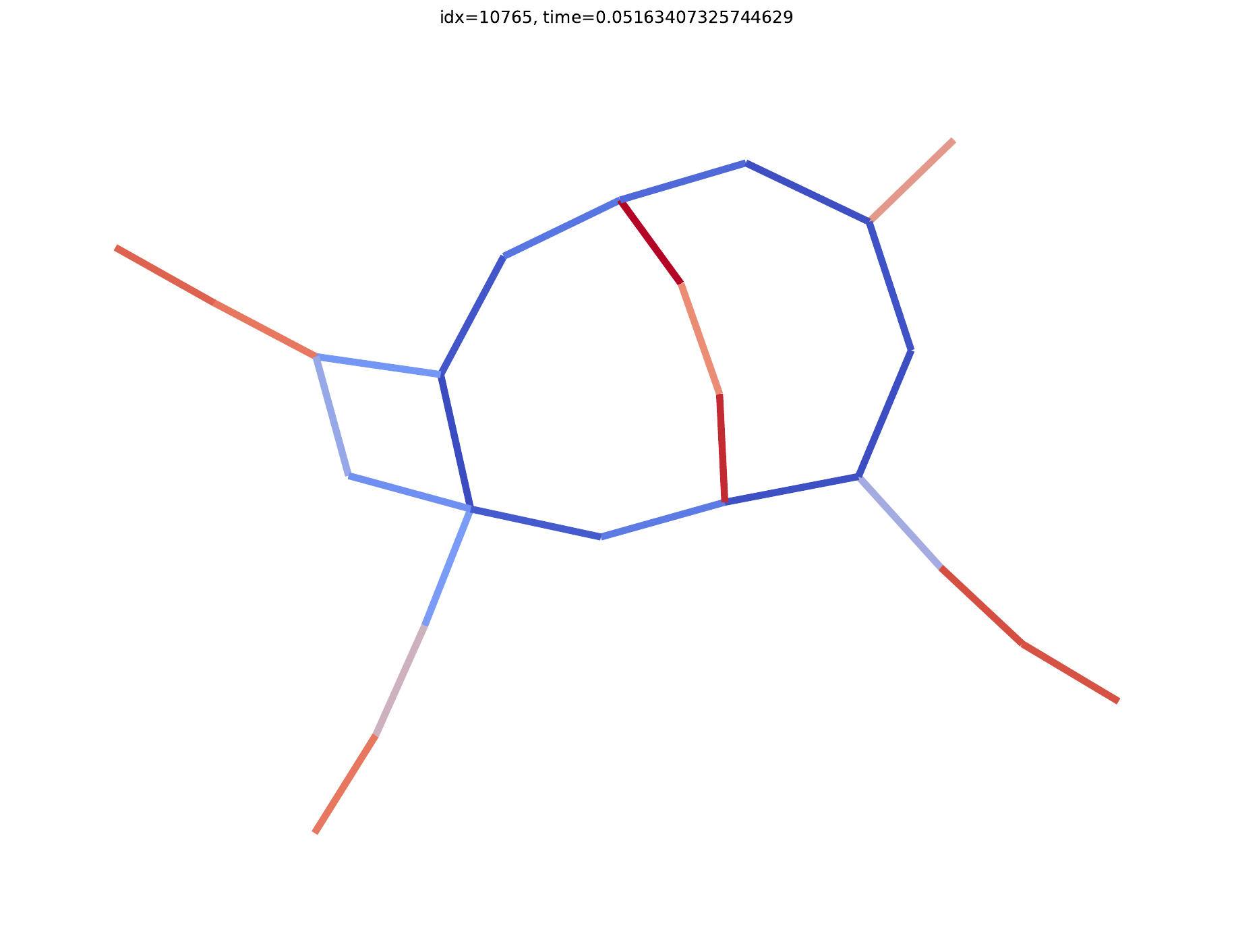} &
\imgcell{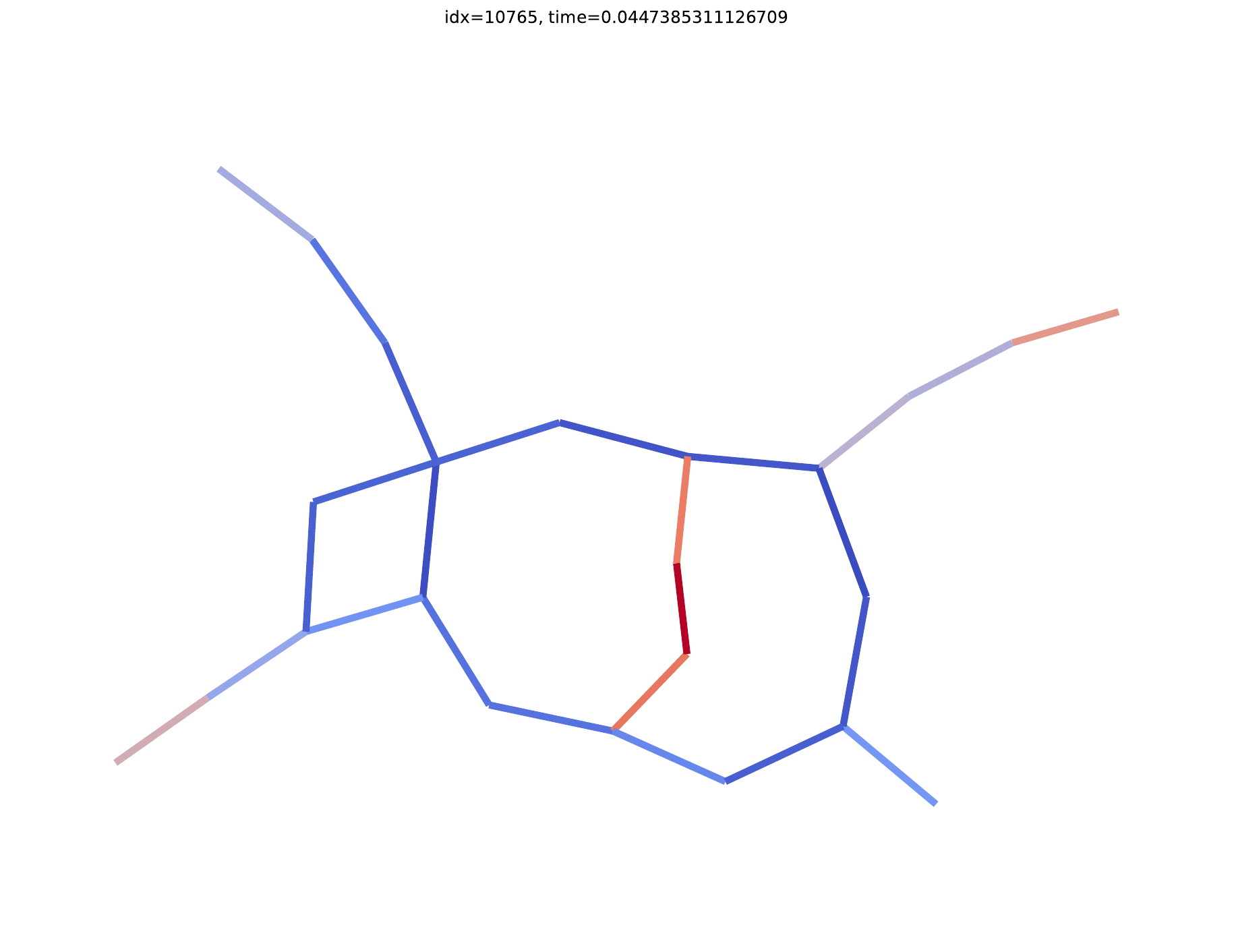} &
\imgcell{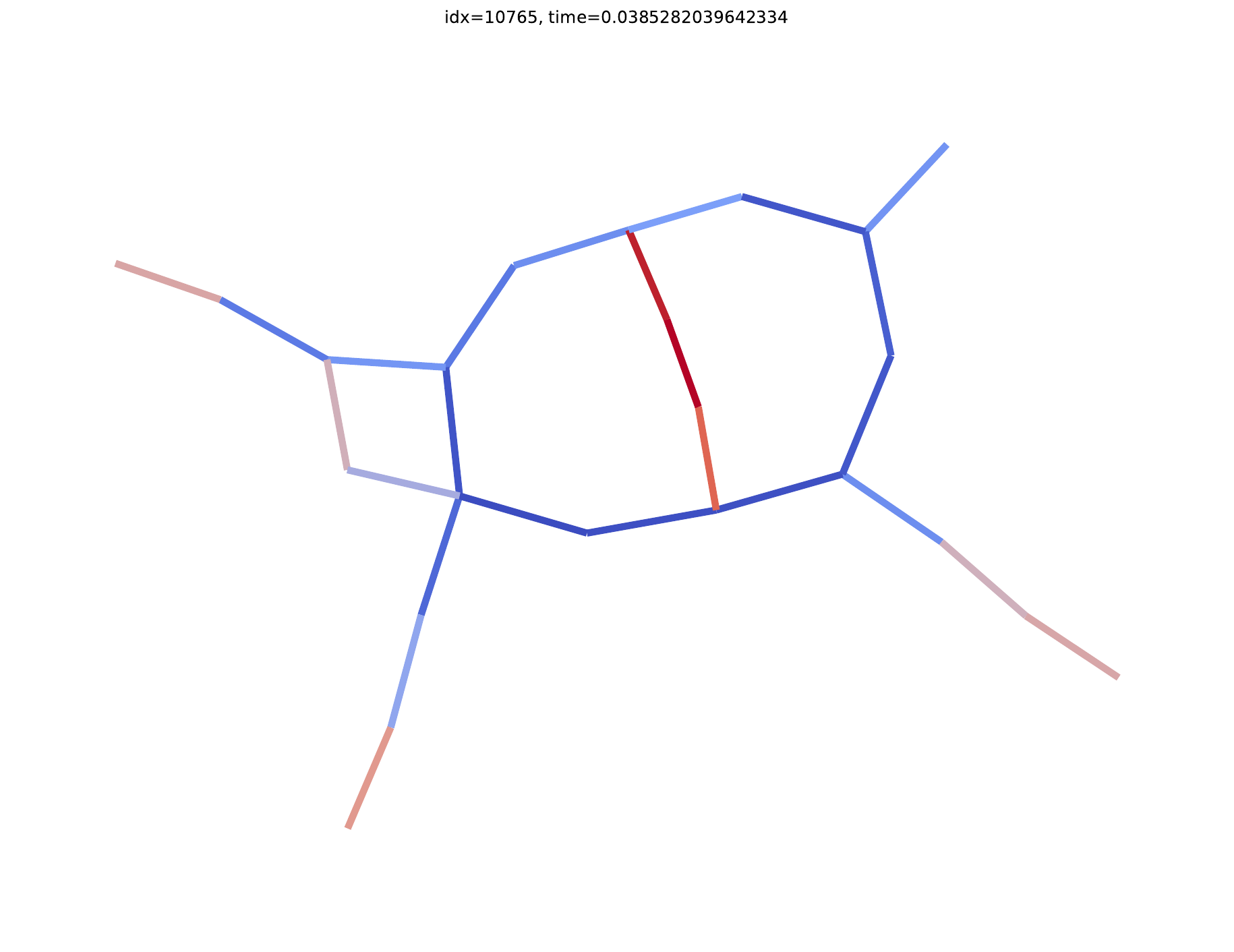} &
\imgcell{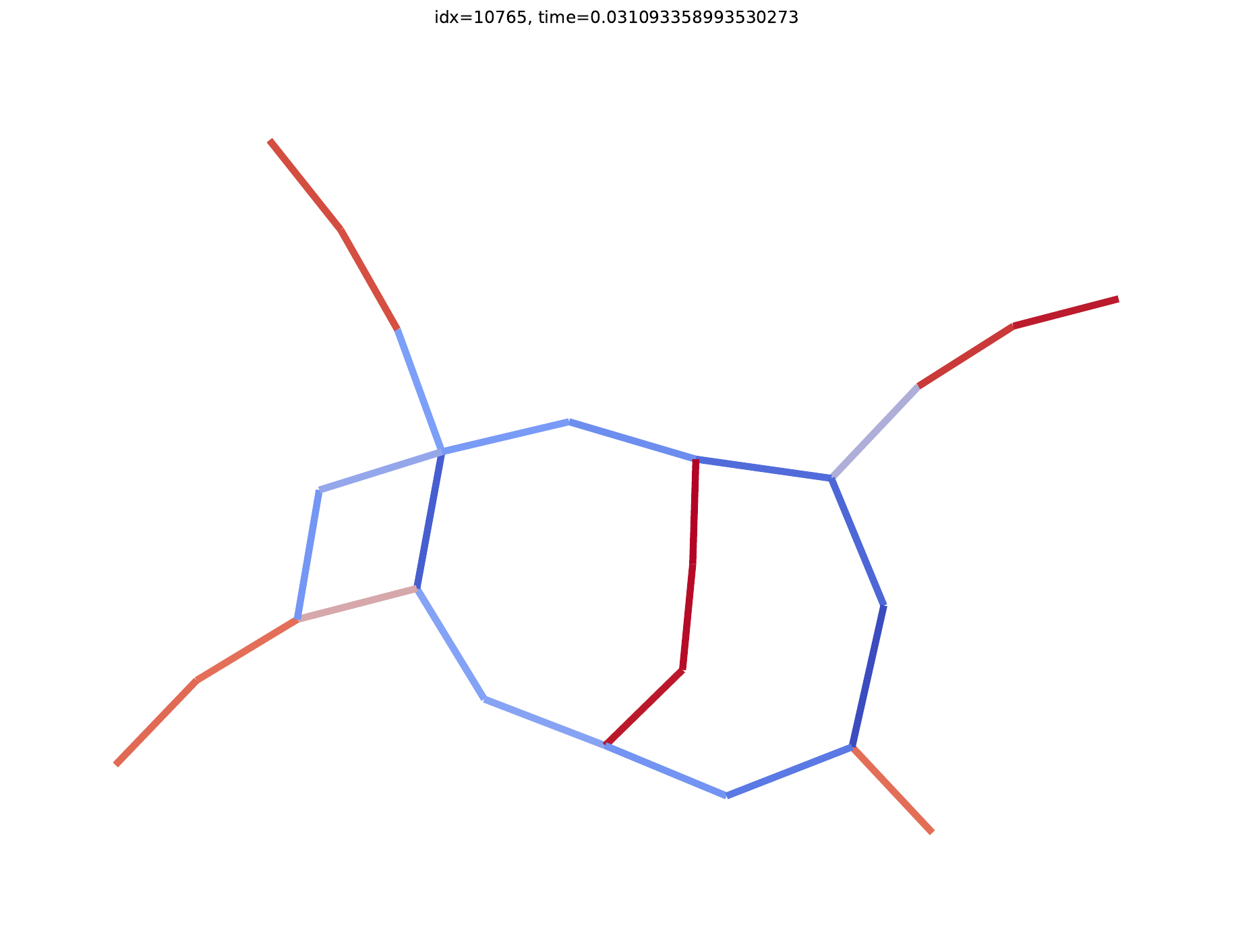} &
\imgcell{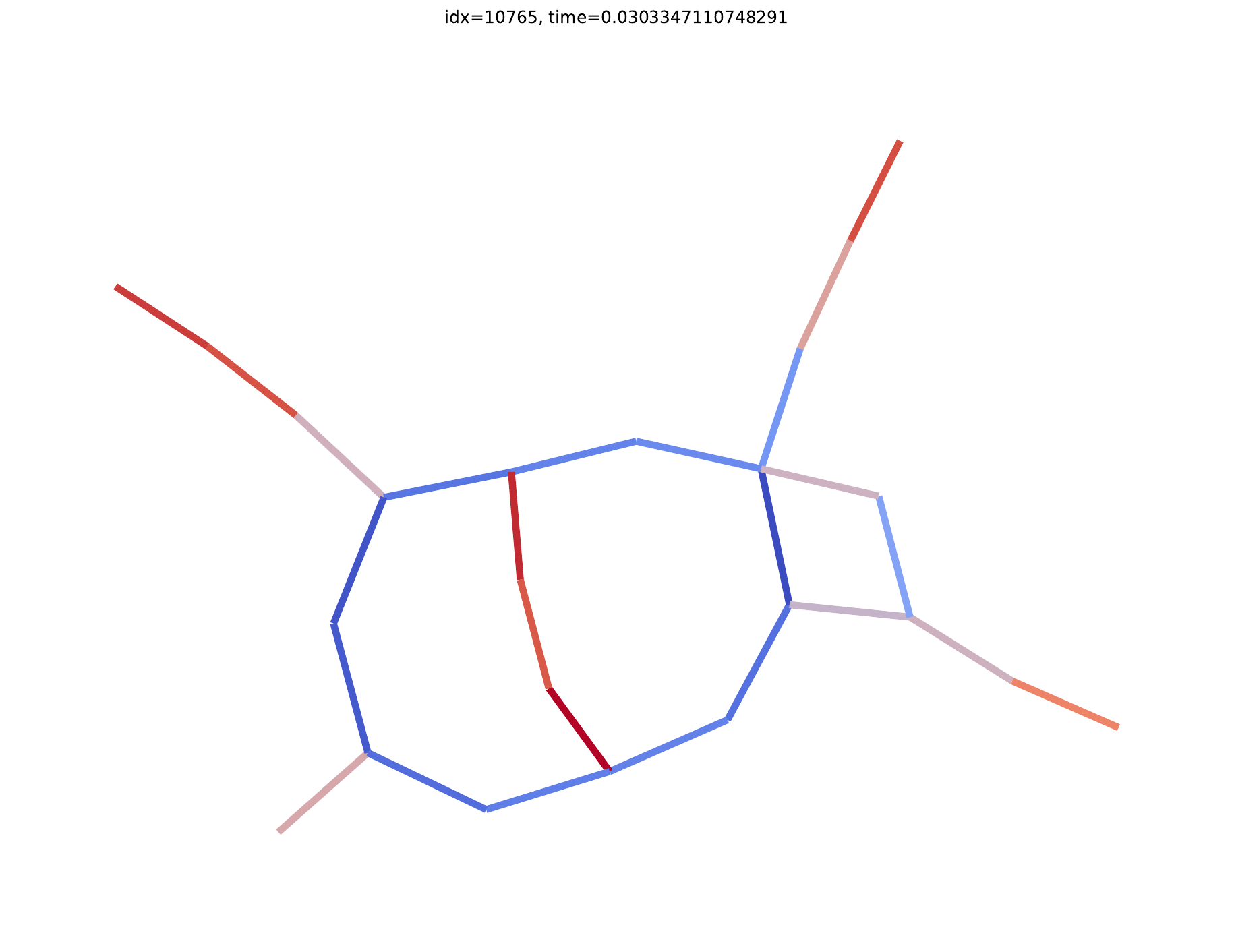} &
\imgcell{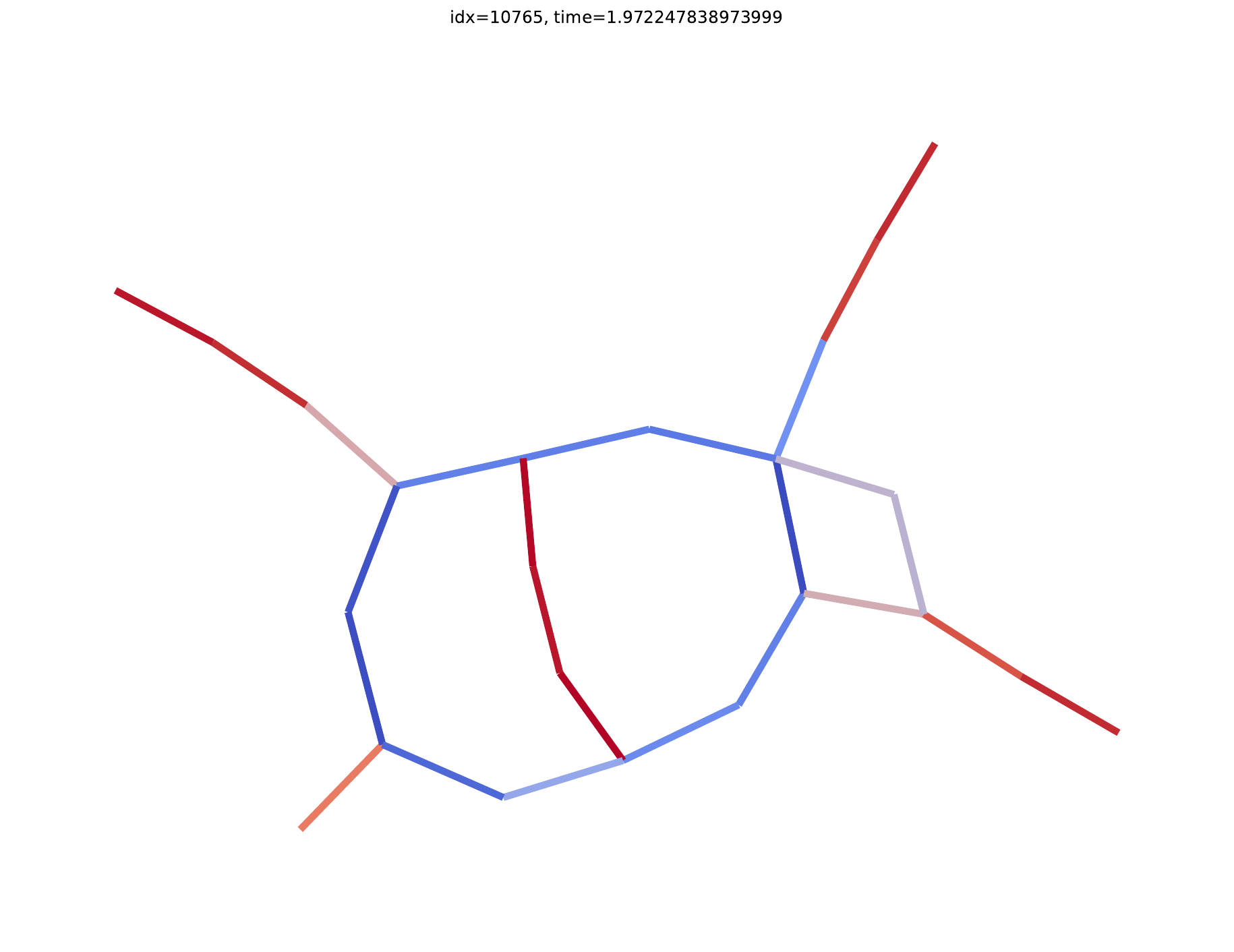} &
\imgcell{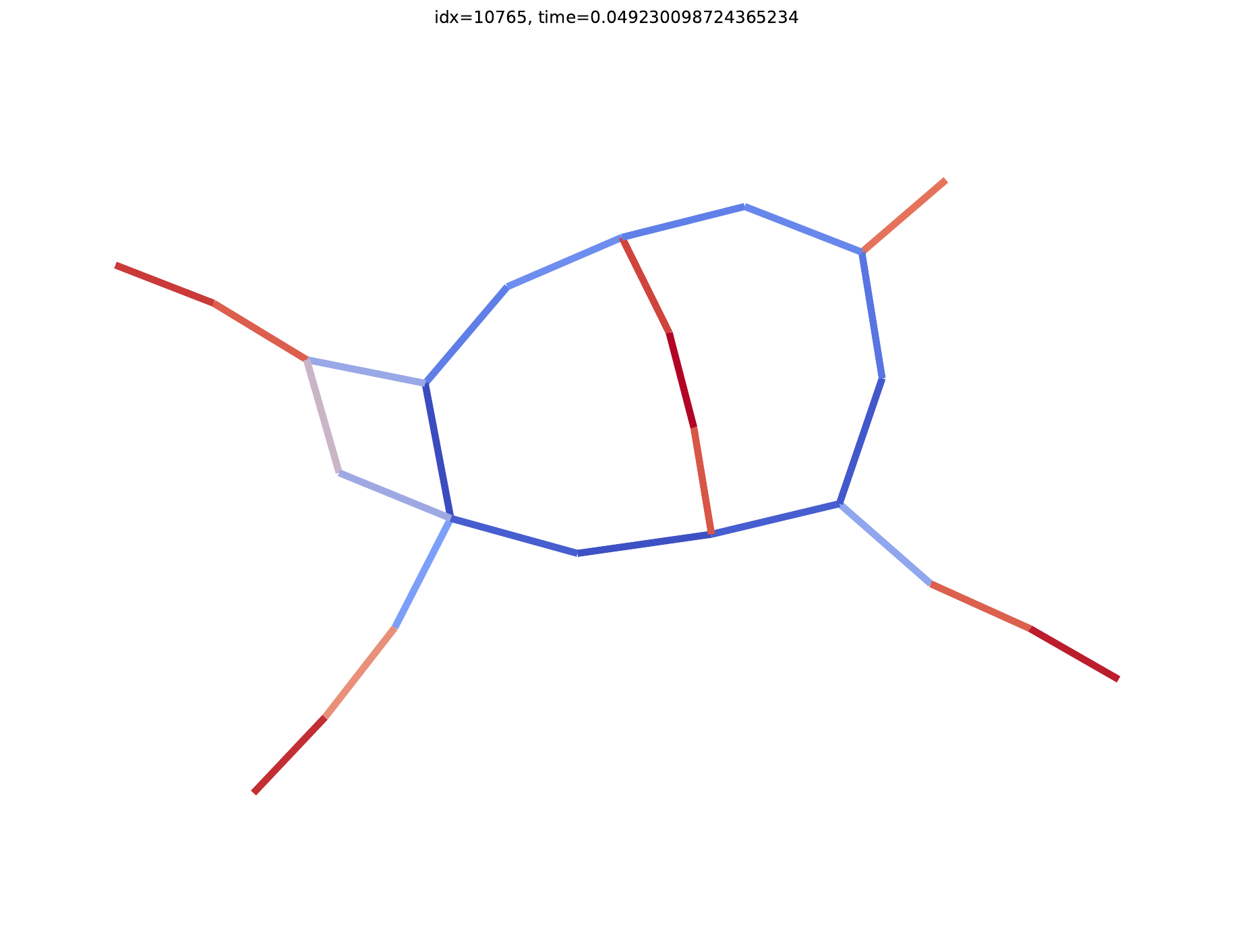} \\

&
t = 0.00s &
t = 0.26s &
t = 0.05s &
t = 0.05s &
t = 90.88s &
t = 0.05s &
t = 0.04s &
t = 0.04s &
t = 0.03s &
t = 0.03s &
t = 0.05s &
t = 0.05s \\

\makecell{\bfseries grafo1936.18\\N = 67\\M = 88} &
\imgcell{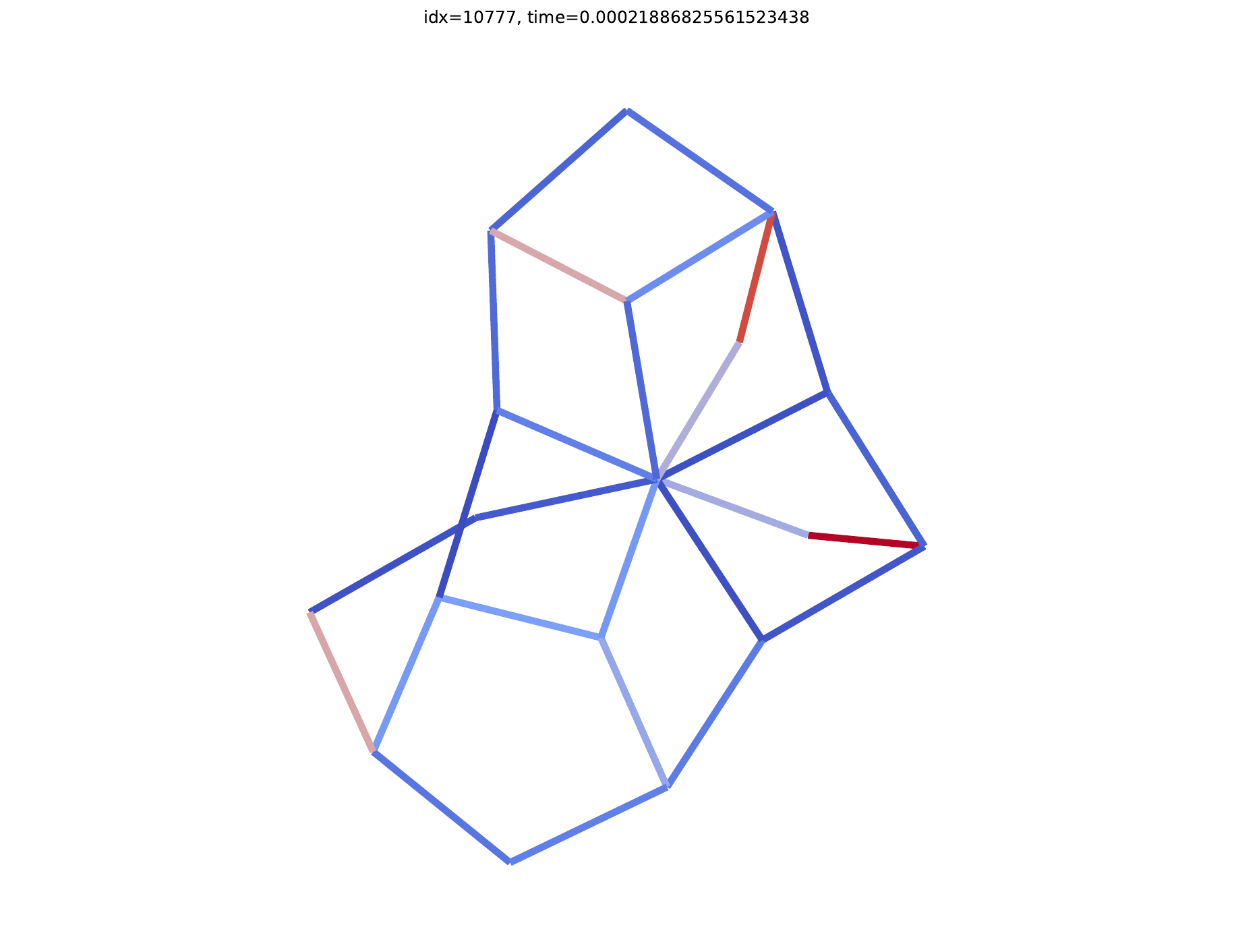} &
\imgcell{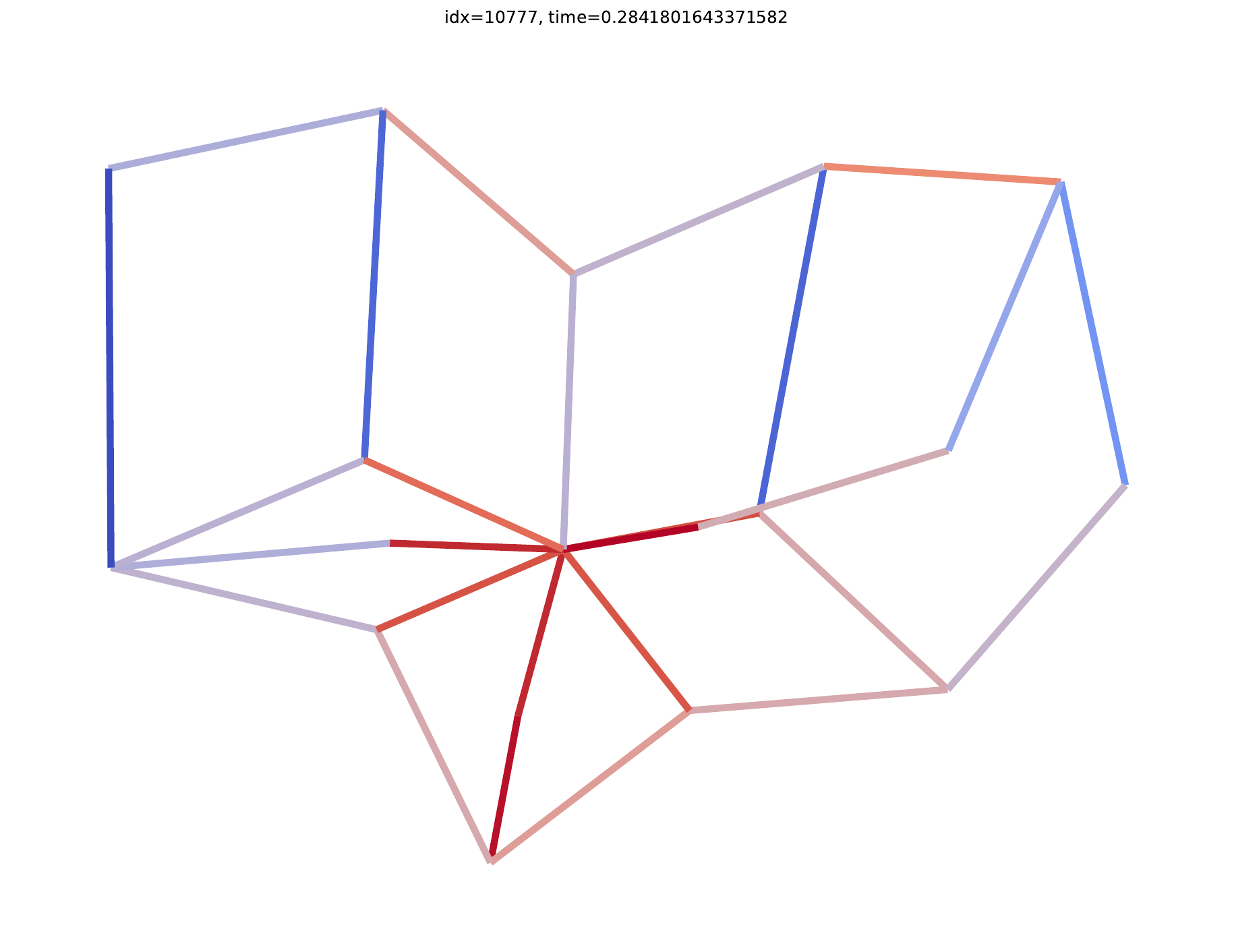} &
\imgcell{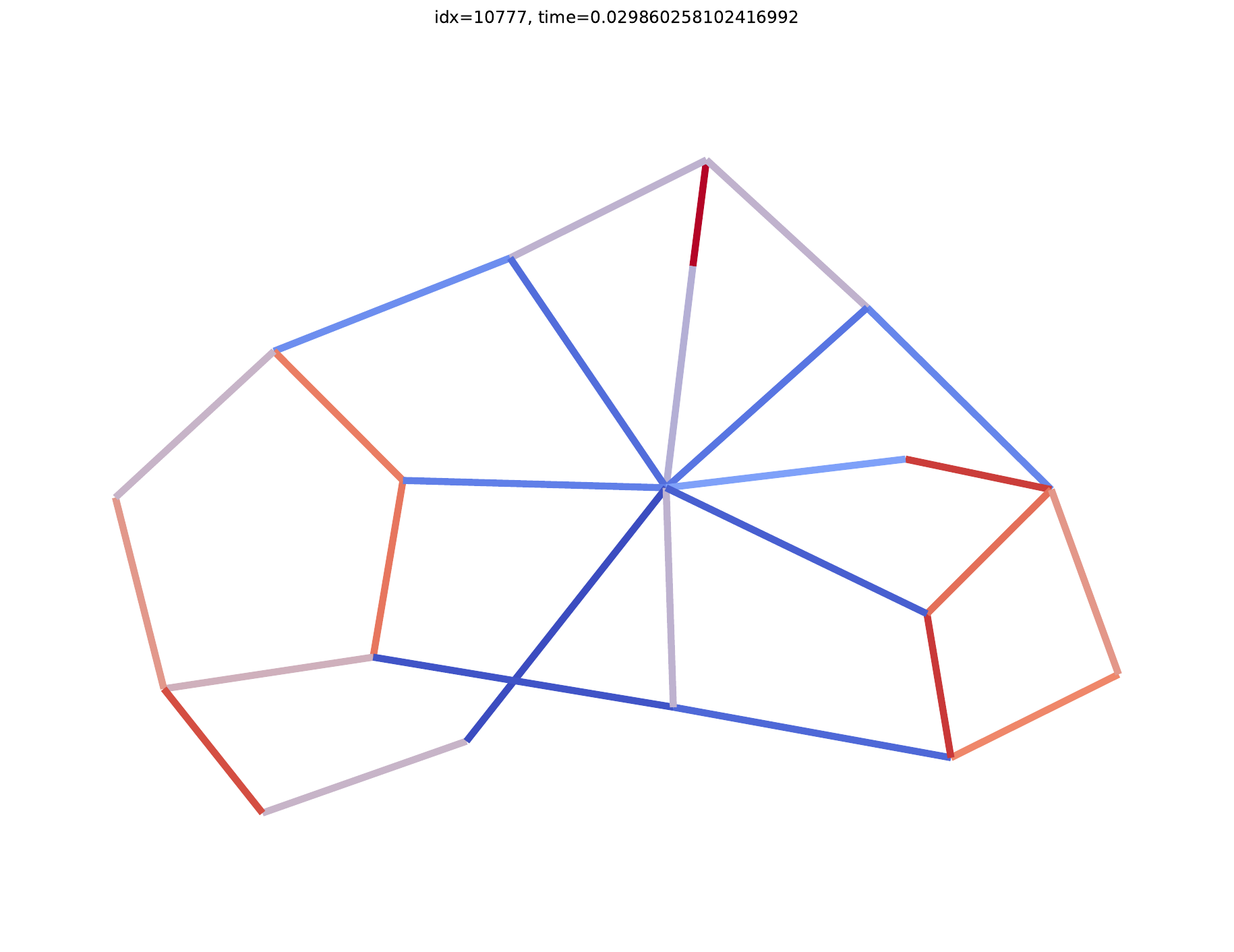} &
\imgcell{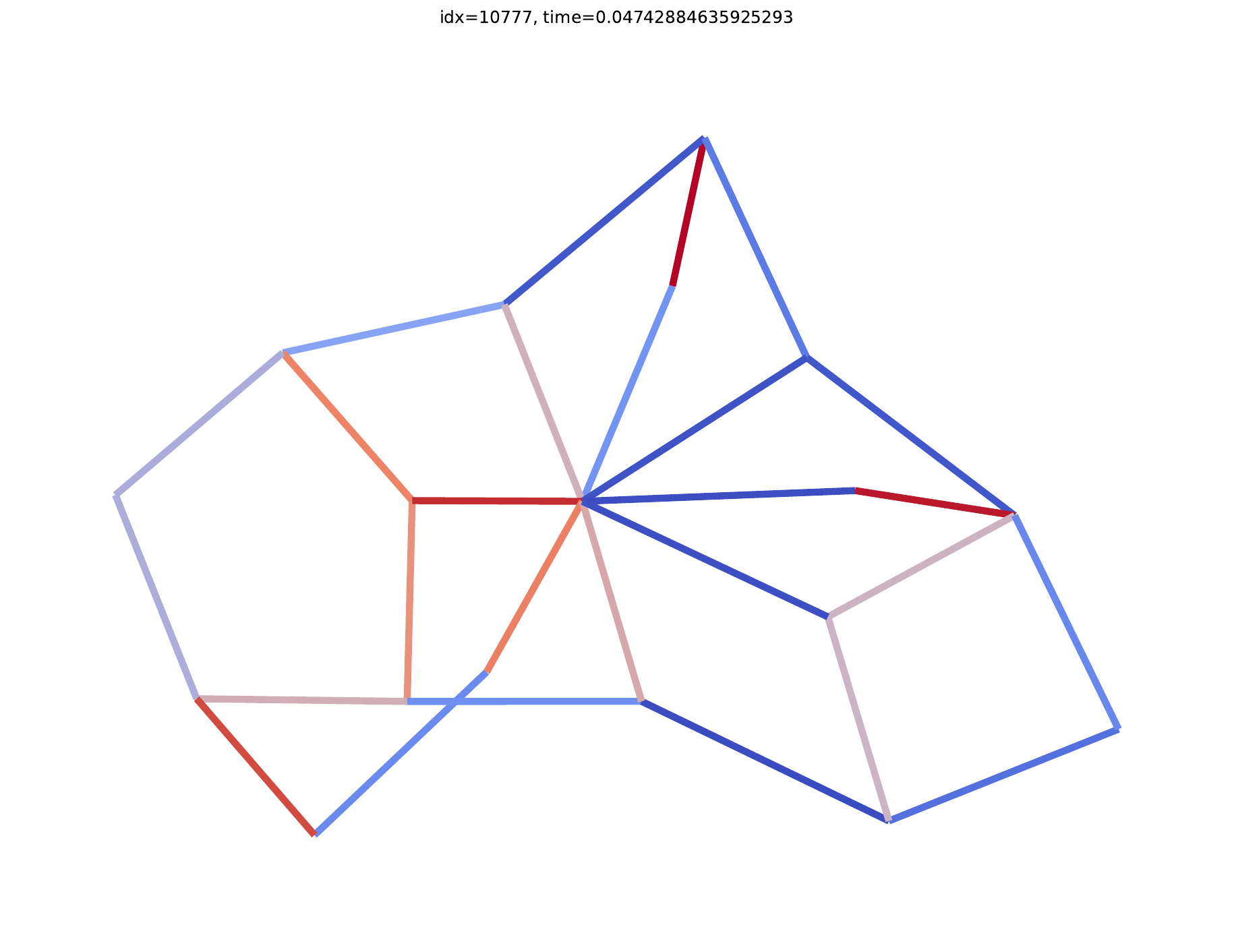} &
\imgcell{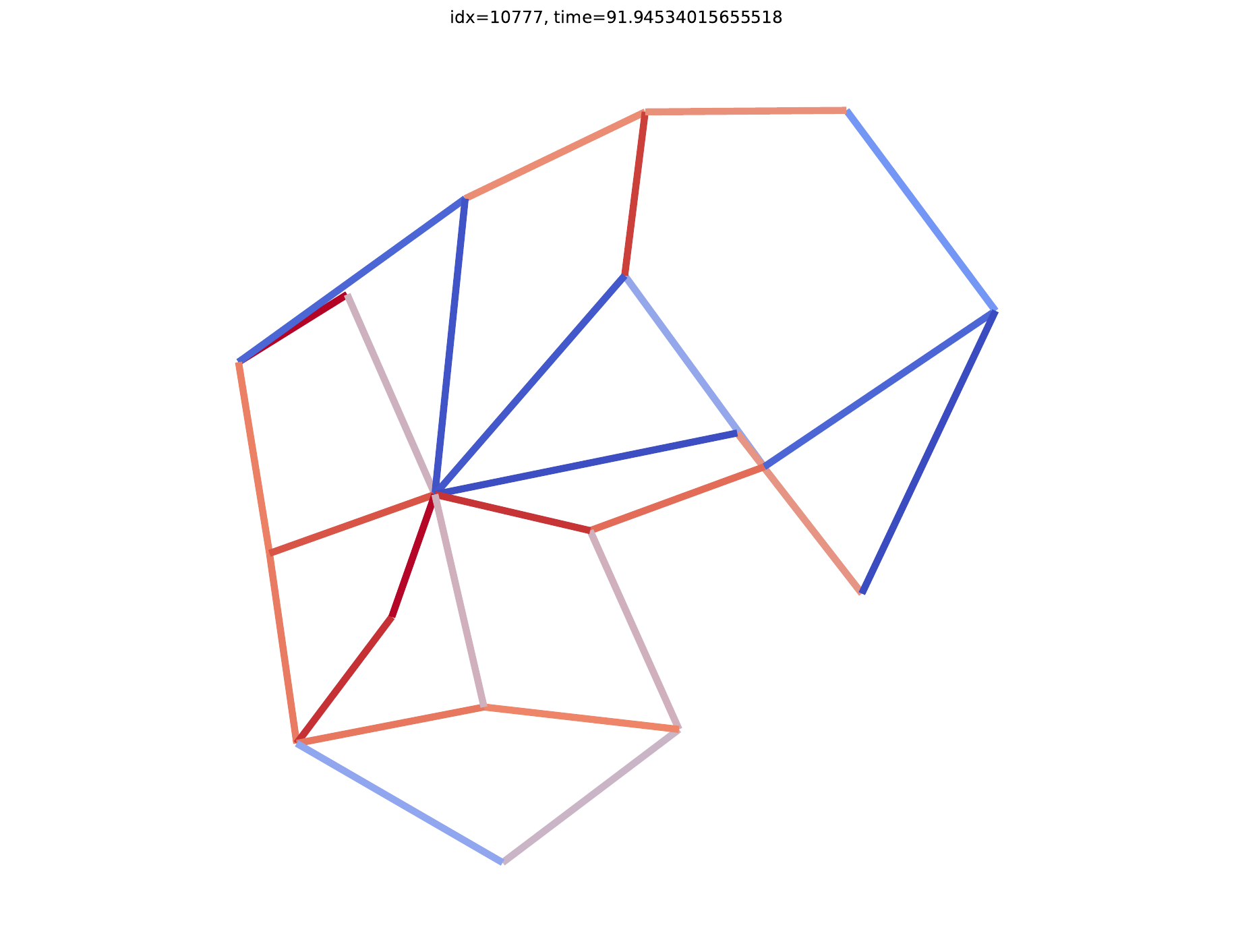} &
\imgcell{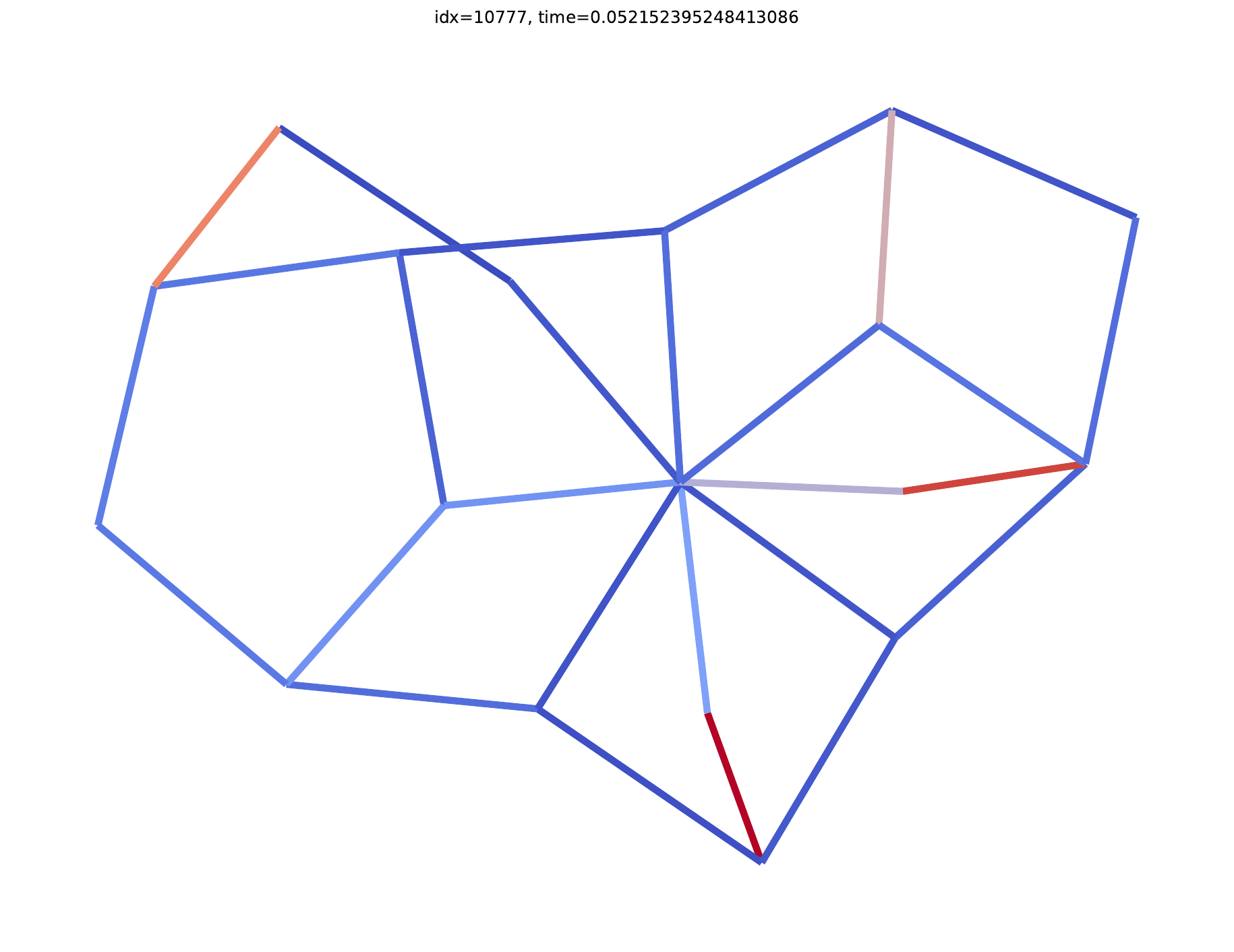} &
\imgcell{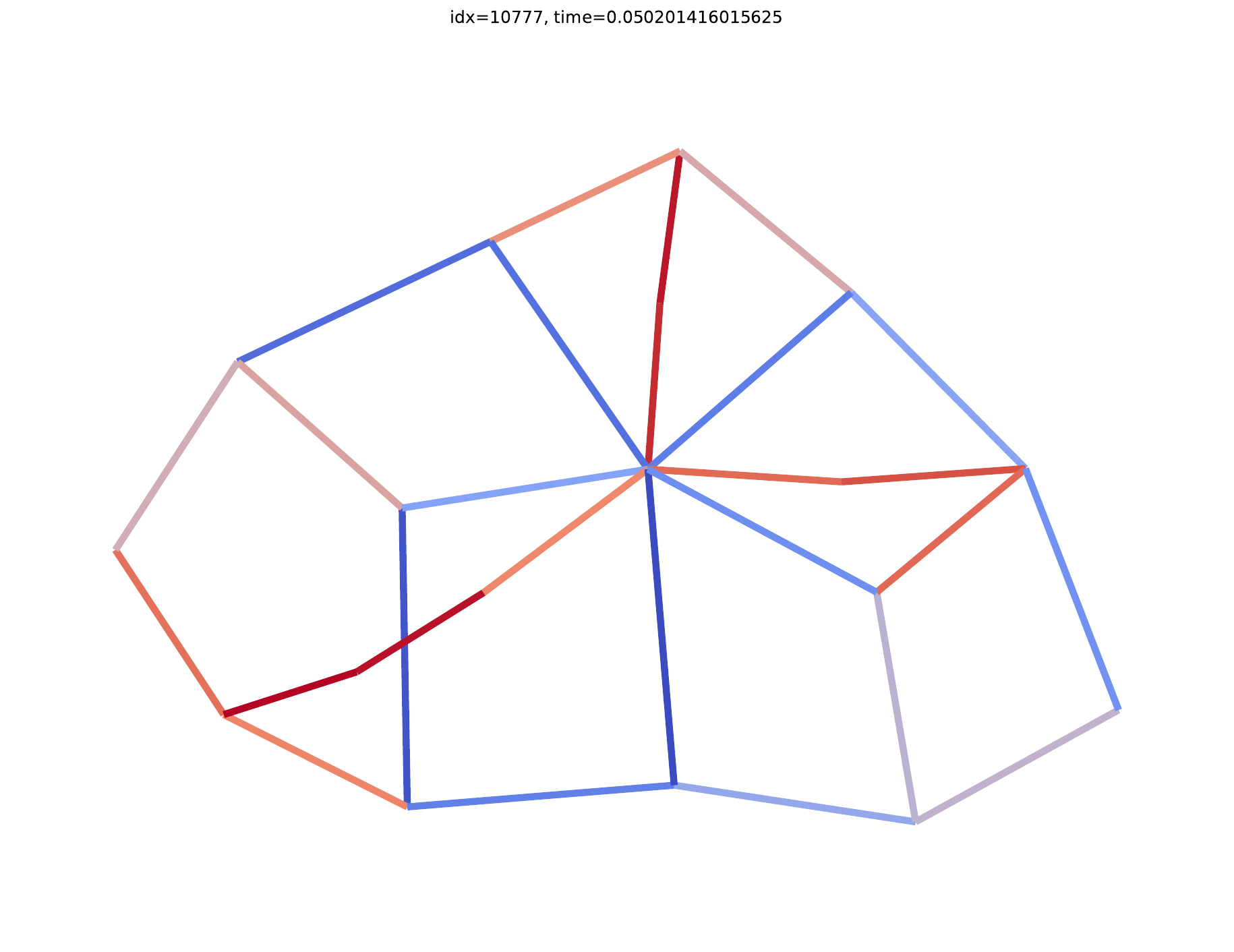} &
\imgcell{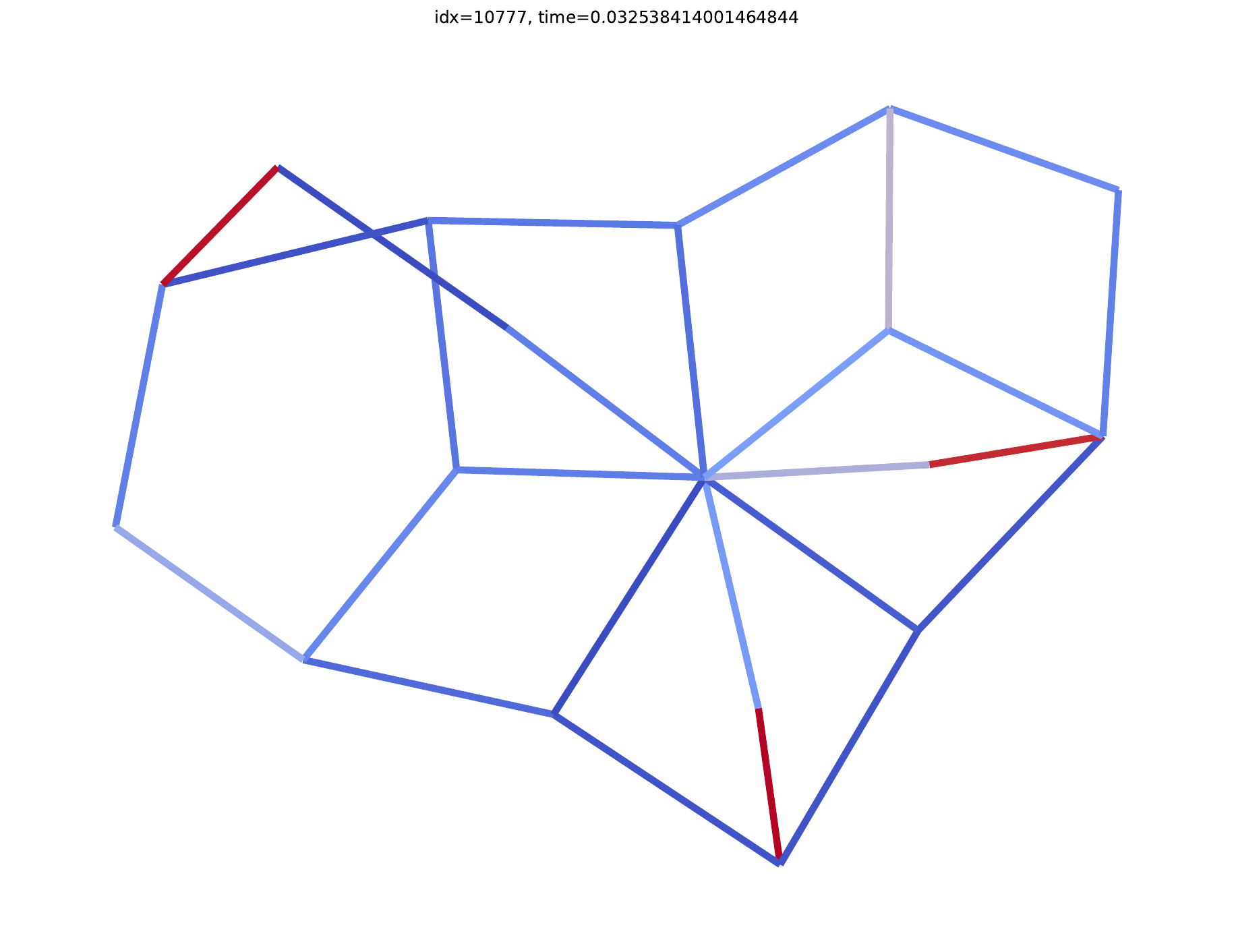} &
\imgcell{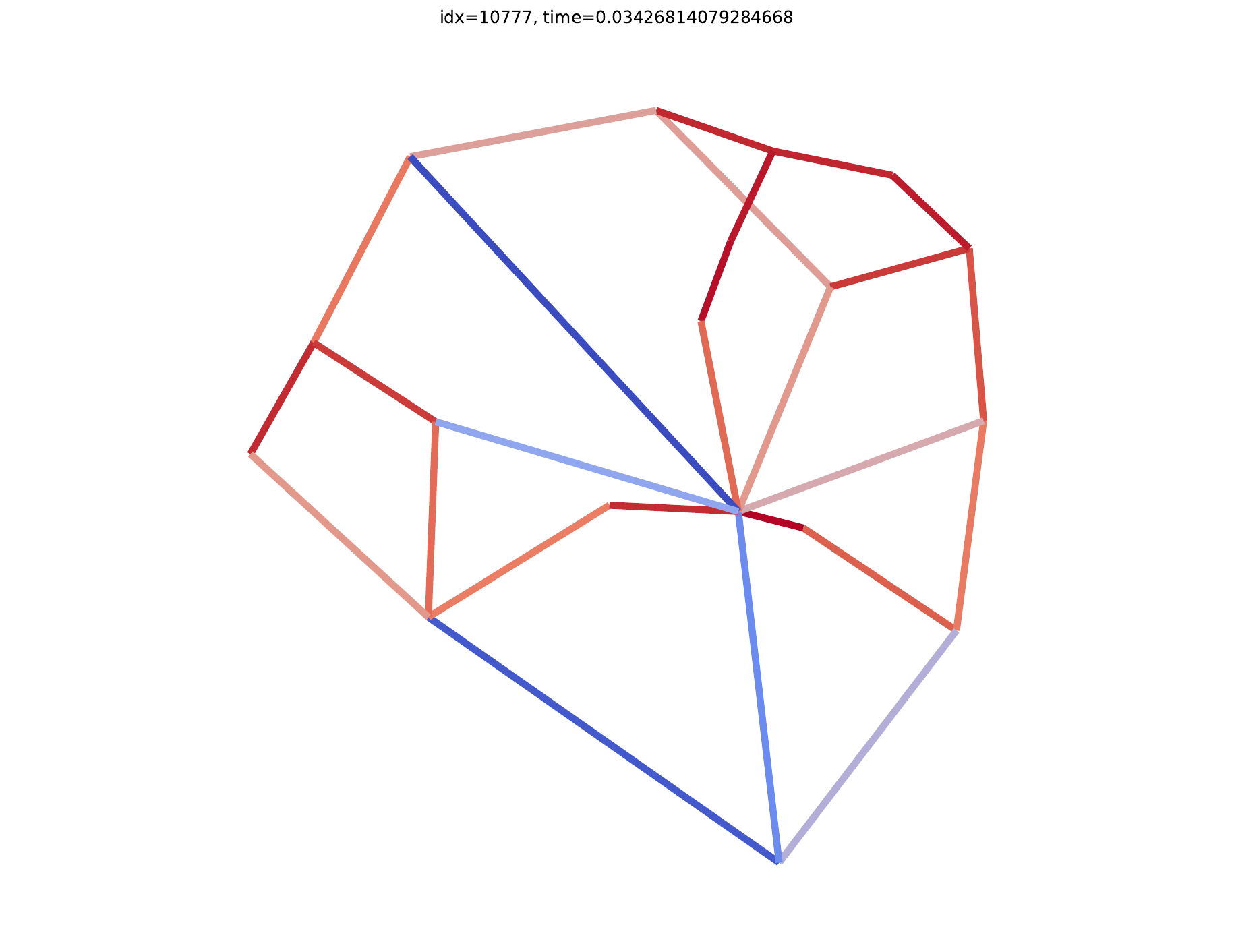} &
\imgcell{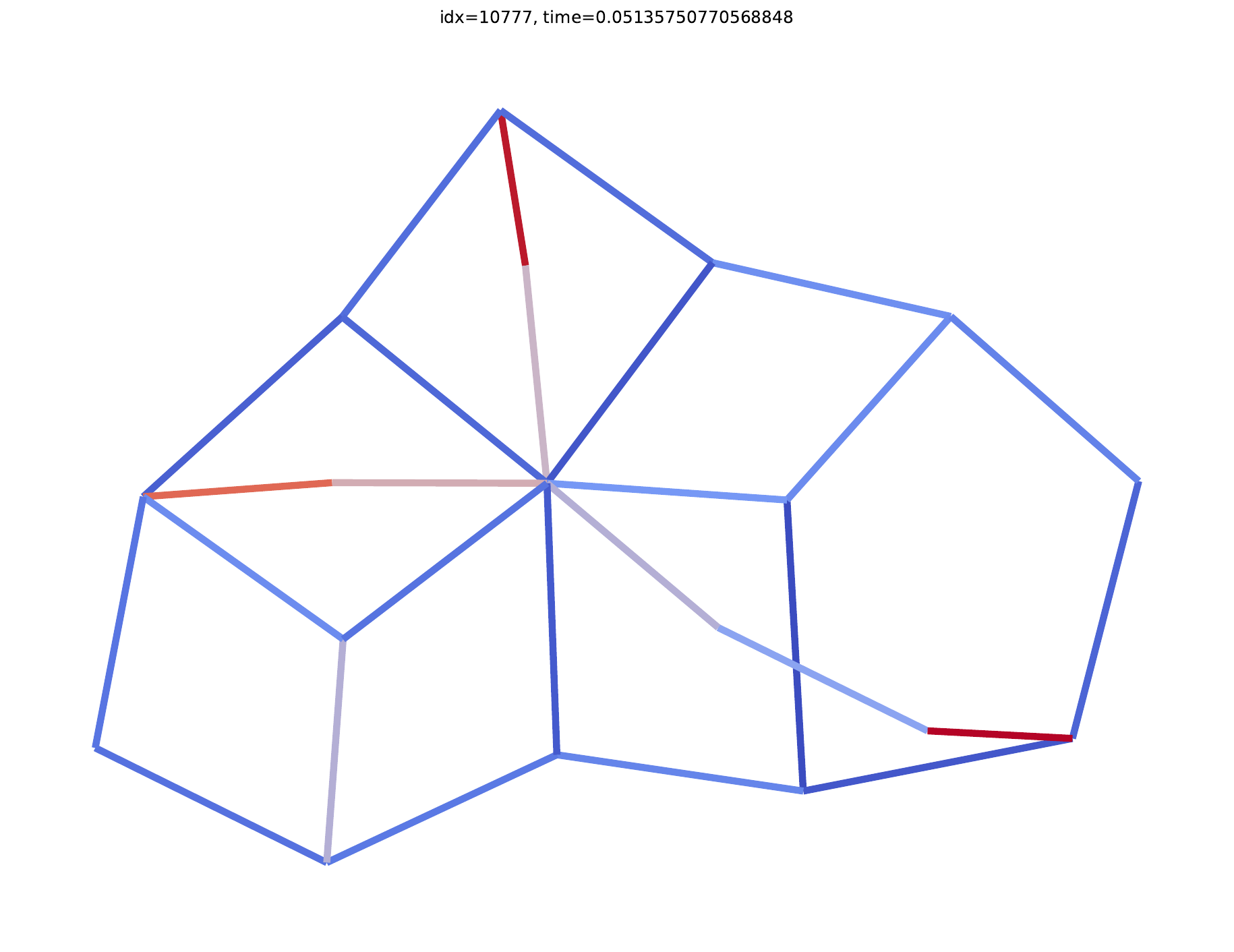} &
\imgcell{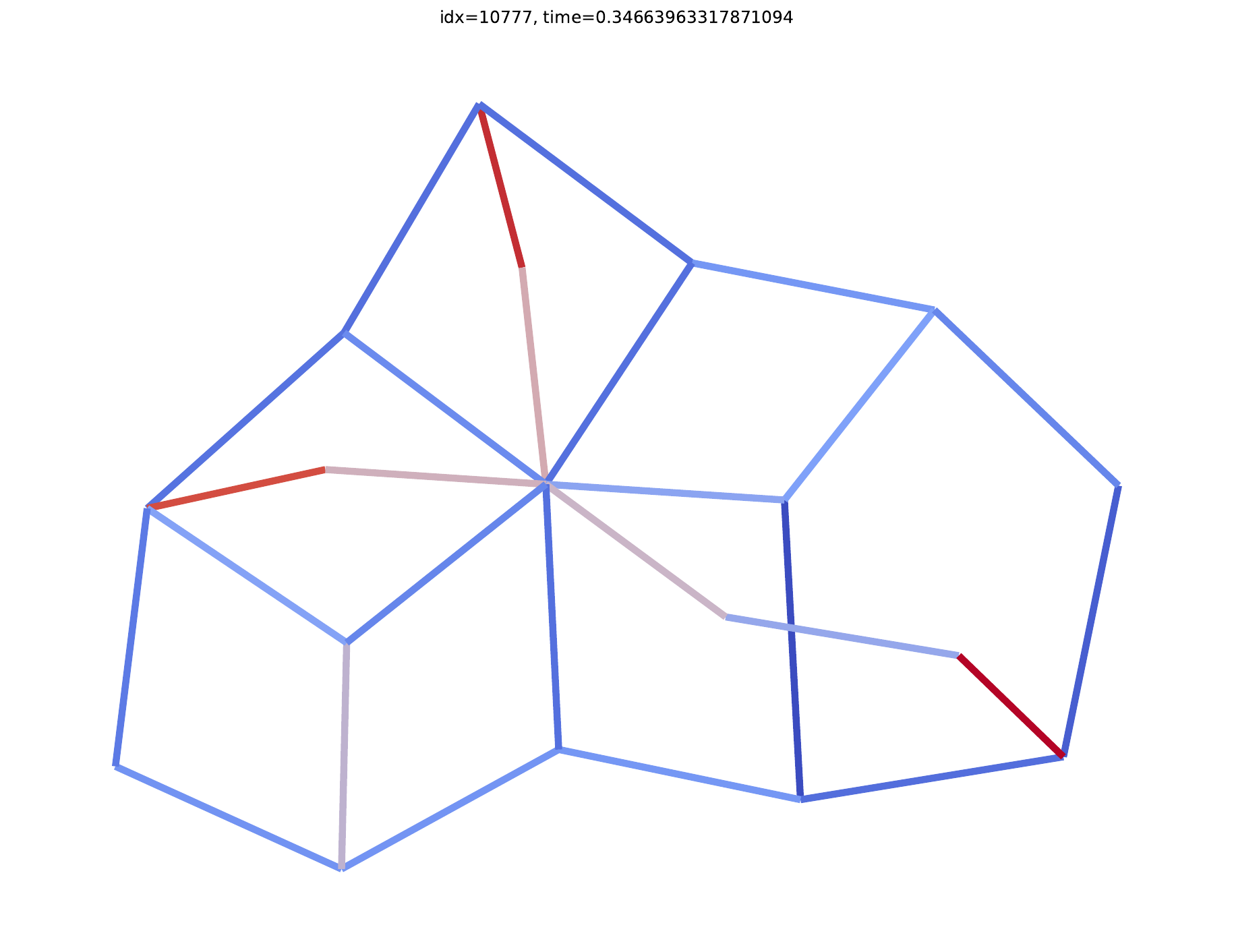} &
\imgcell{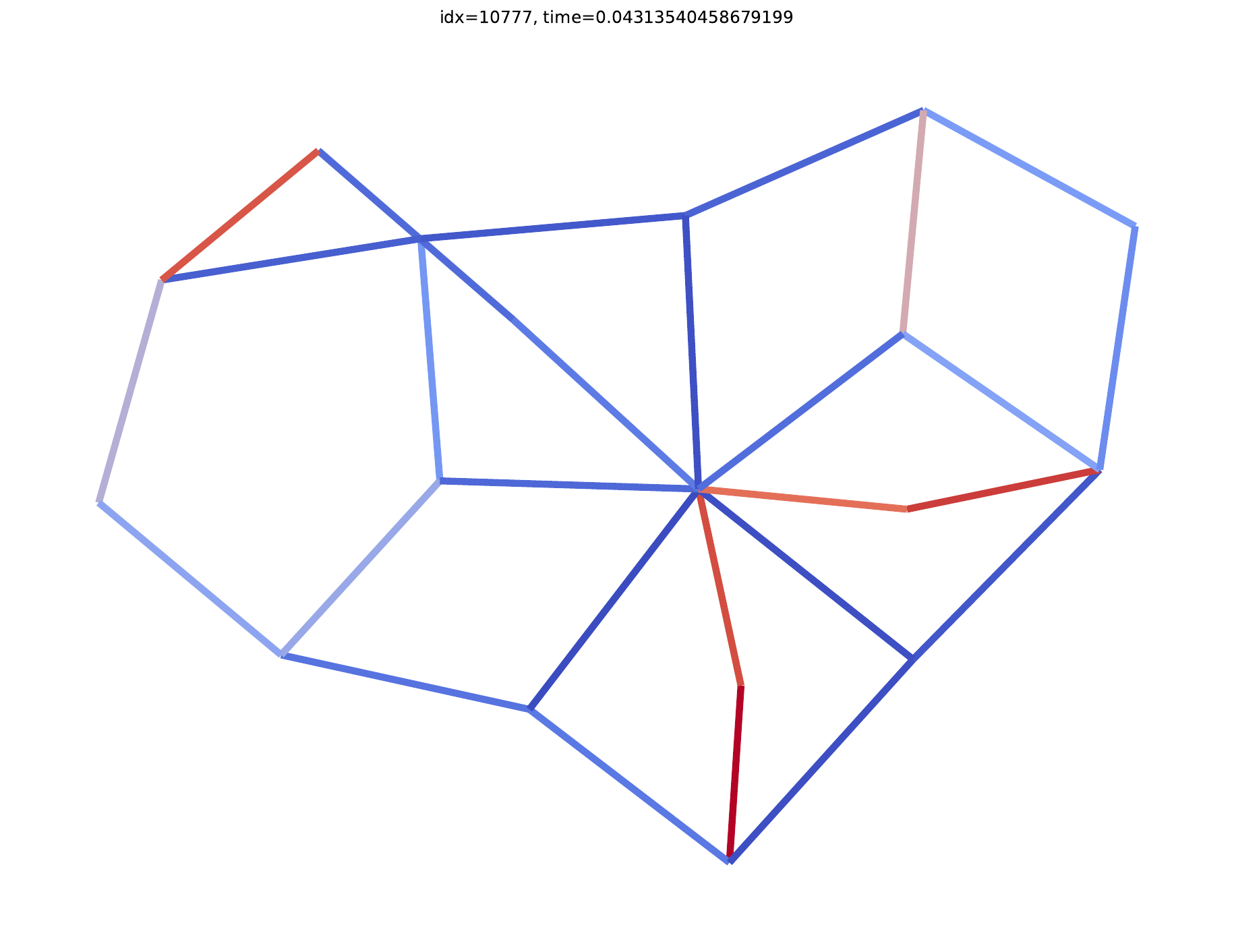} \\

&
t = 0.00s &
t = 0.28s &
t = 0.03s &
t = 0.05s &
t = 91.95s &
t = 0.05s &
t = 0.05s &
t = 0.05s &
t = 0.03s &
t = 0.05s &
t = 0.05s &
t = 0.04s \\

\makecell{\bfseries grafo9711.66\\N = 31\\M = 37} &
\imgcell{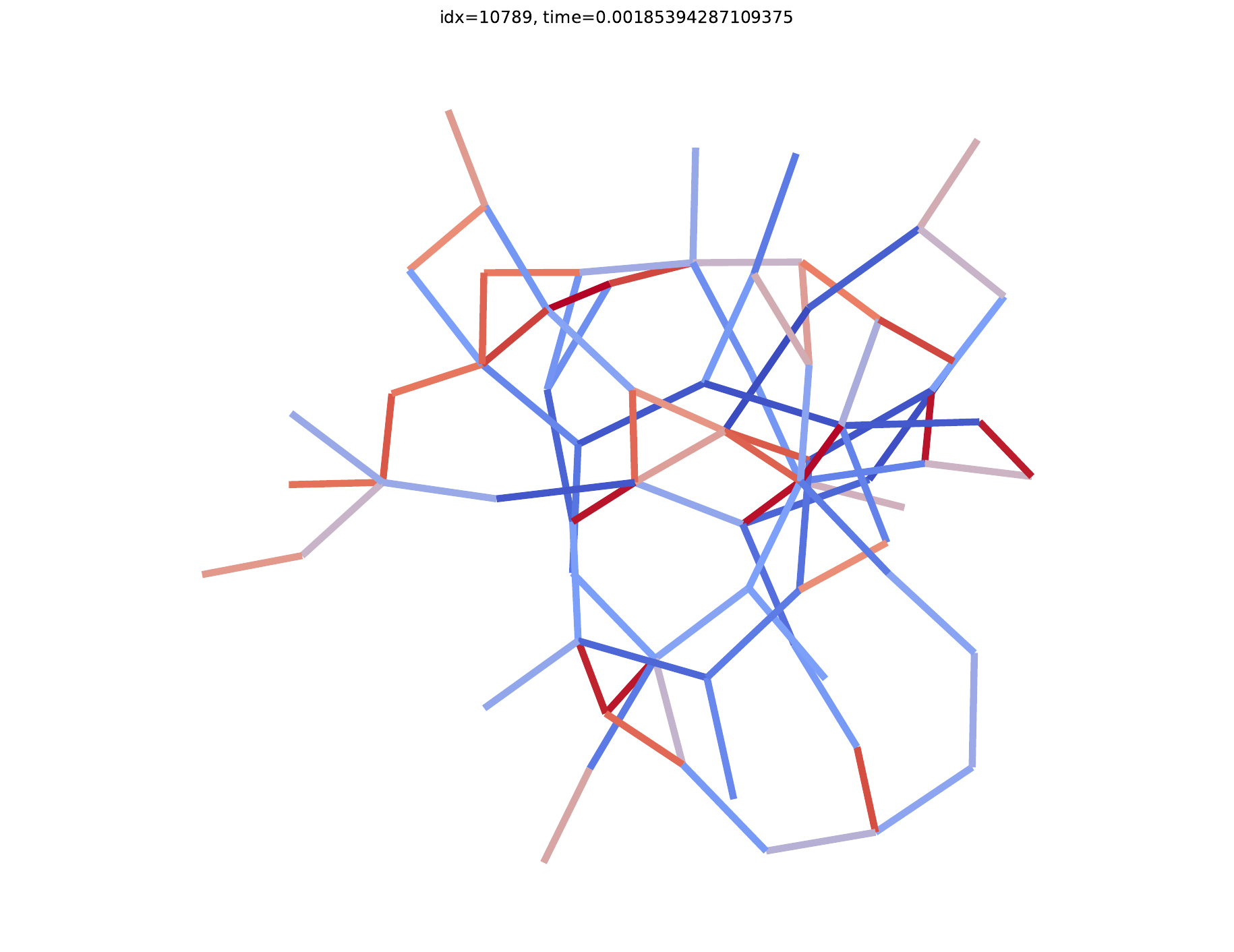} &
\imgcell{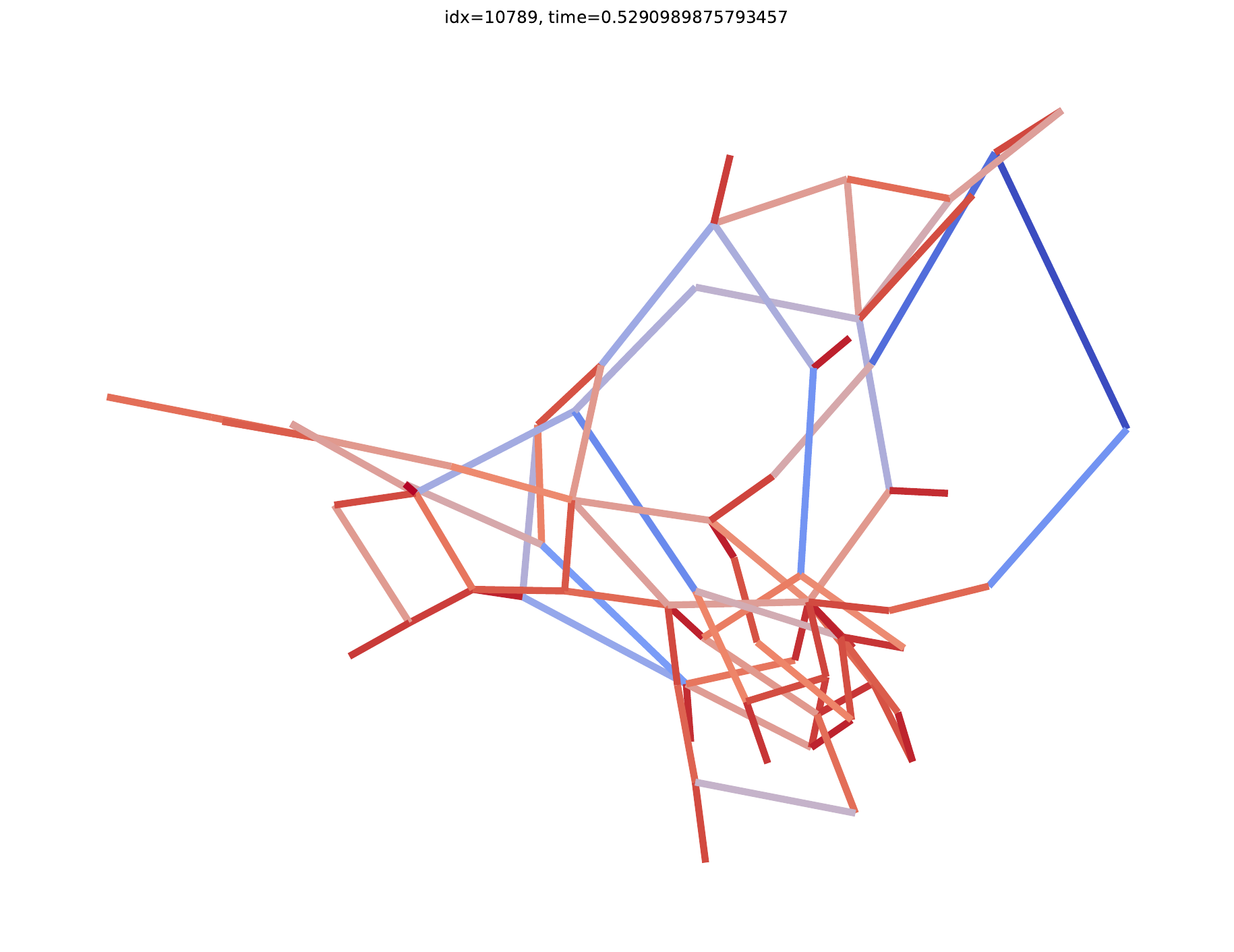} &
\imgcell{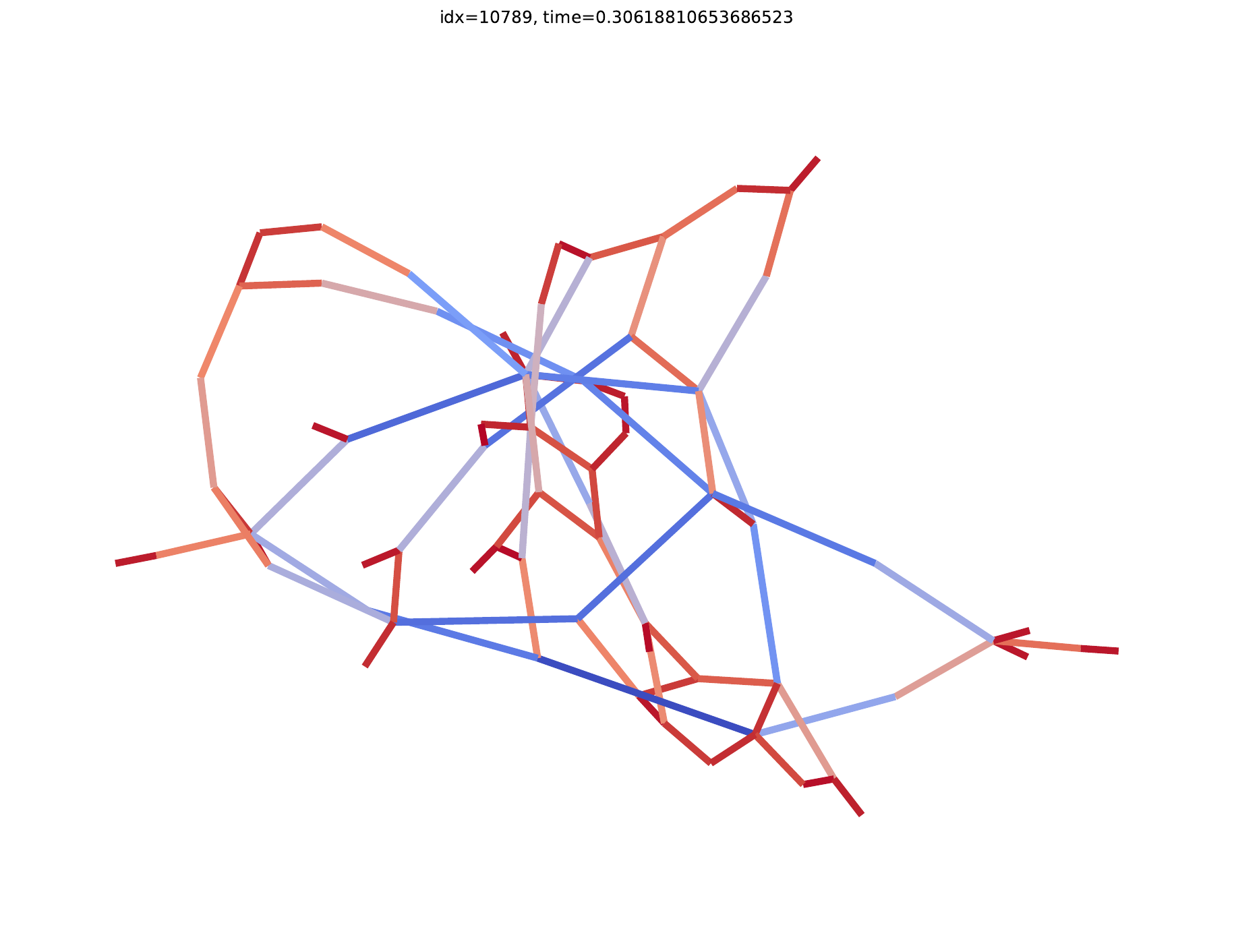} &
\imgcell{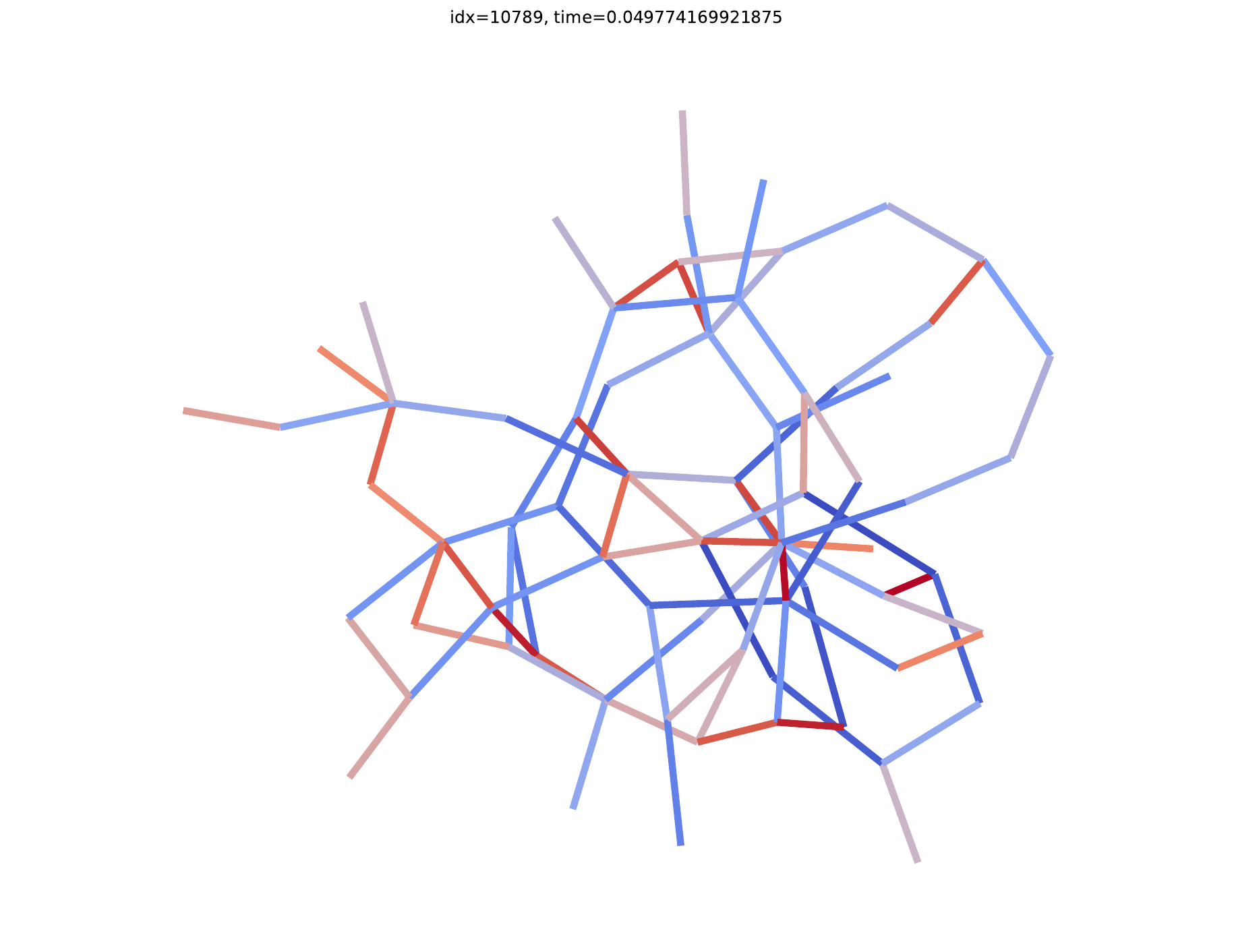} &
\imgcell{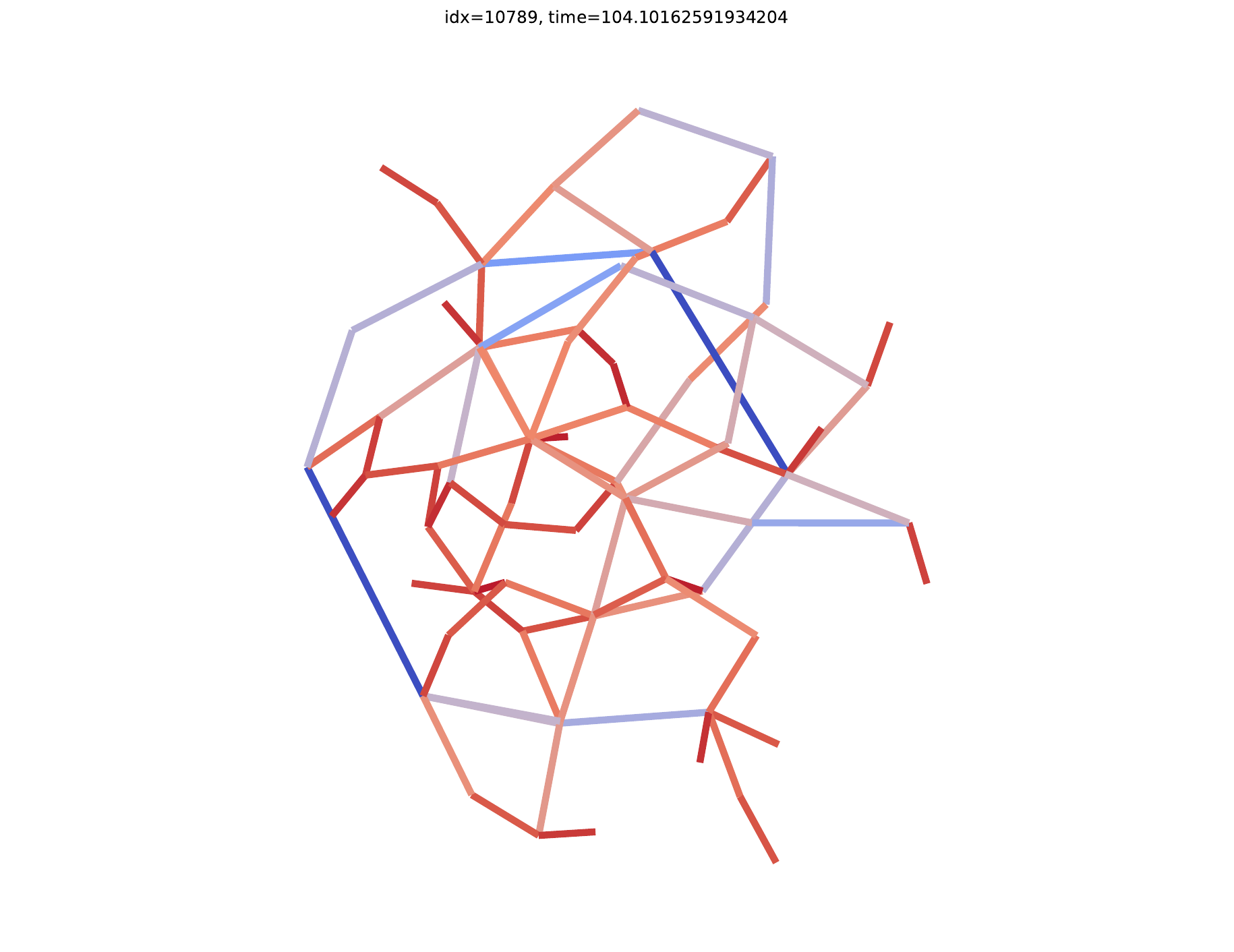} &
\imgcell{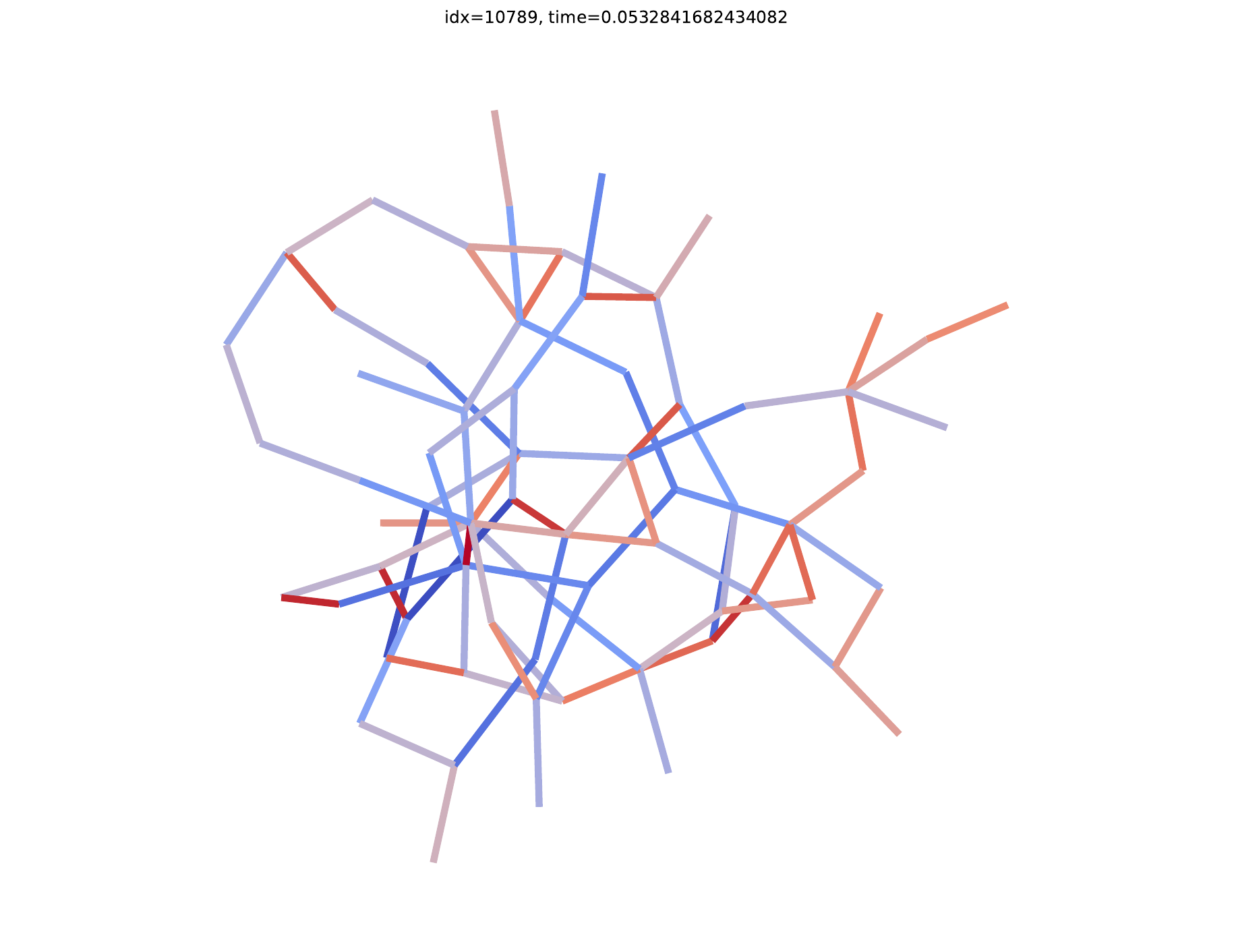} &
\imgcell{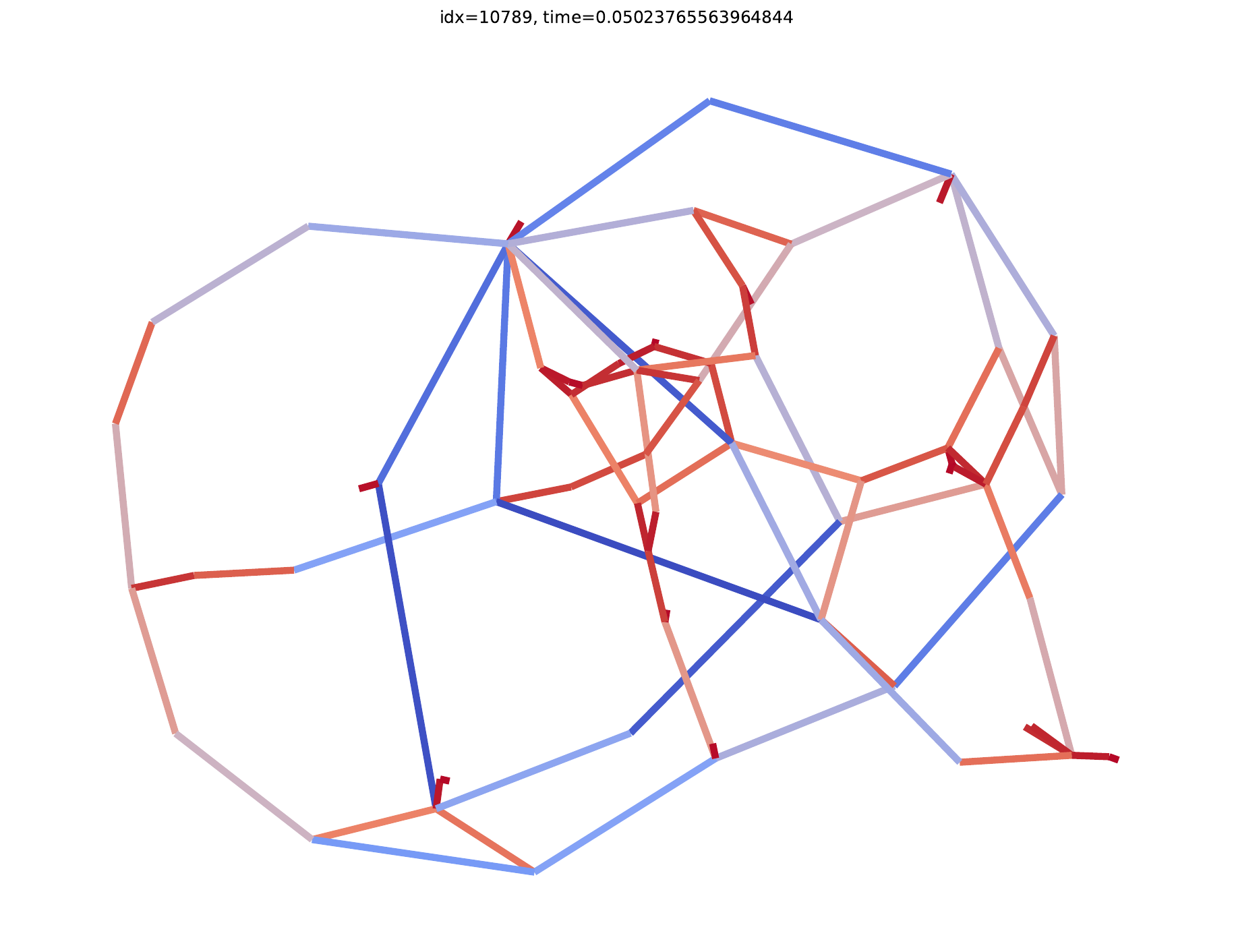} &
\imgcell{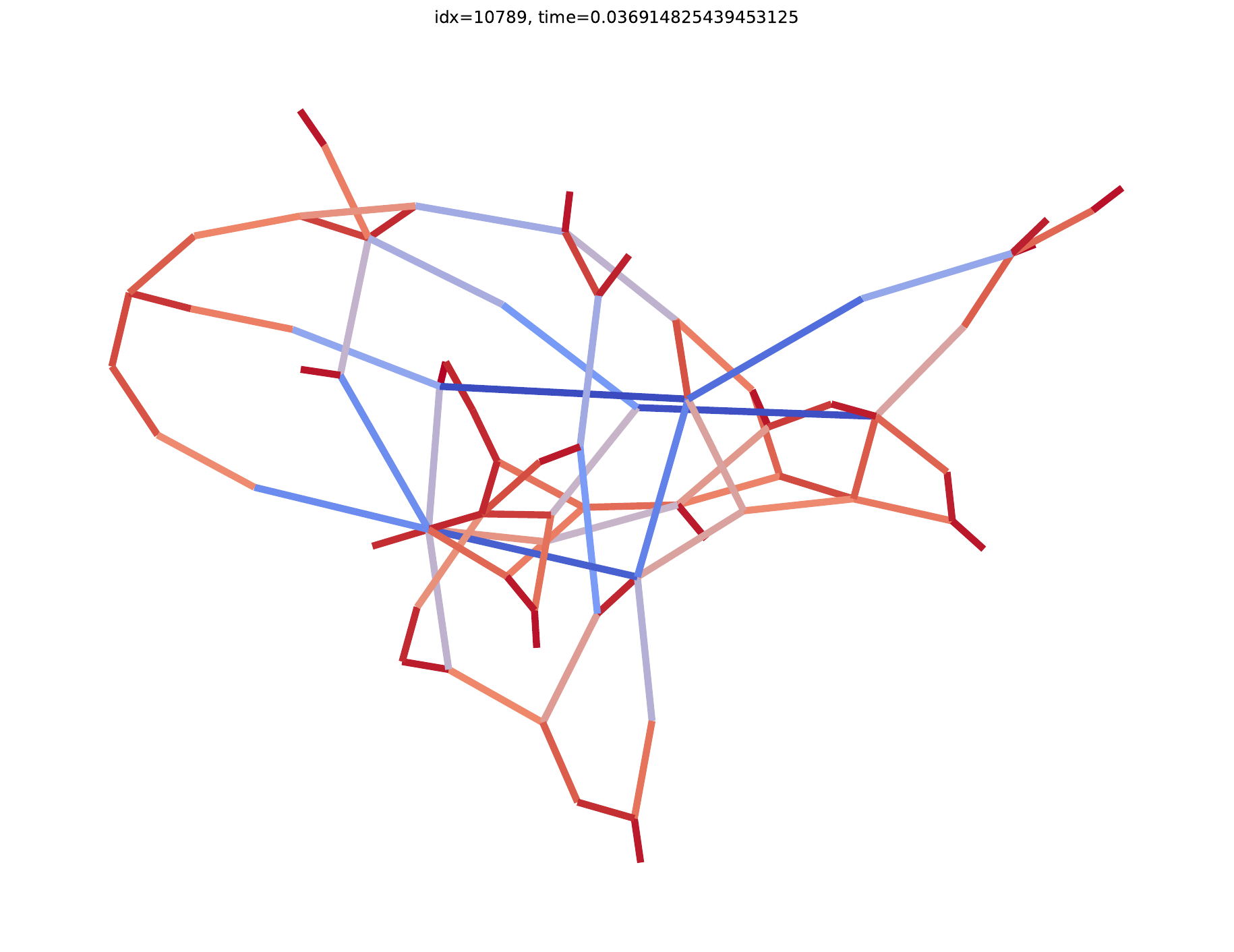} &
\imgcell{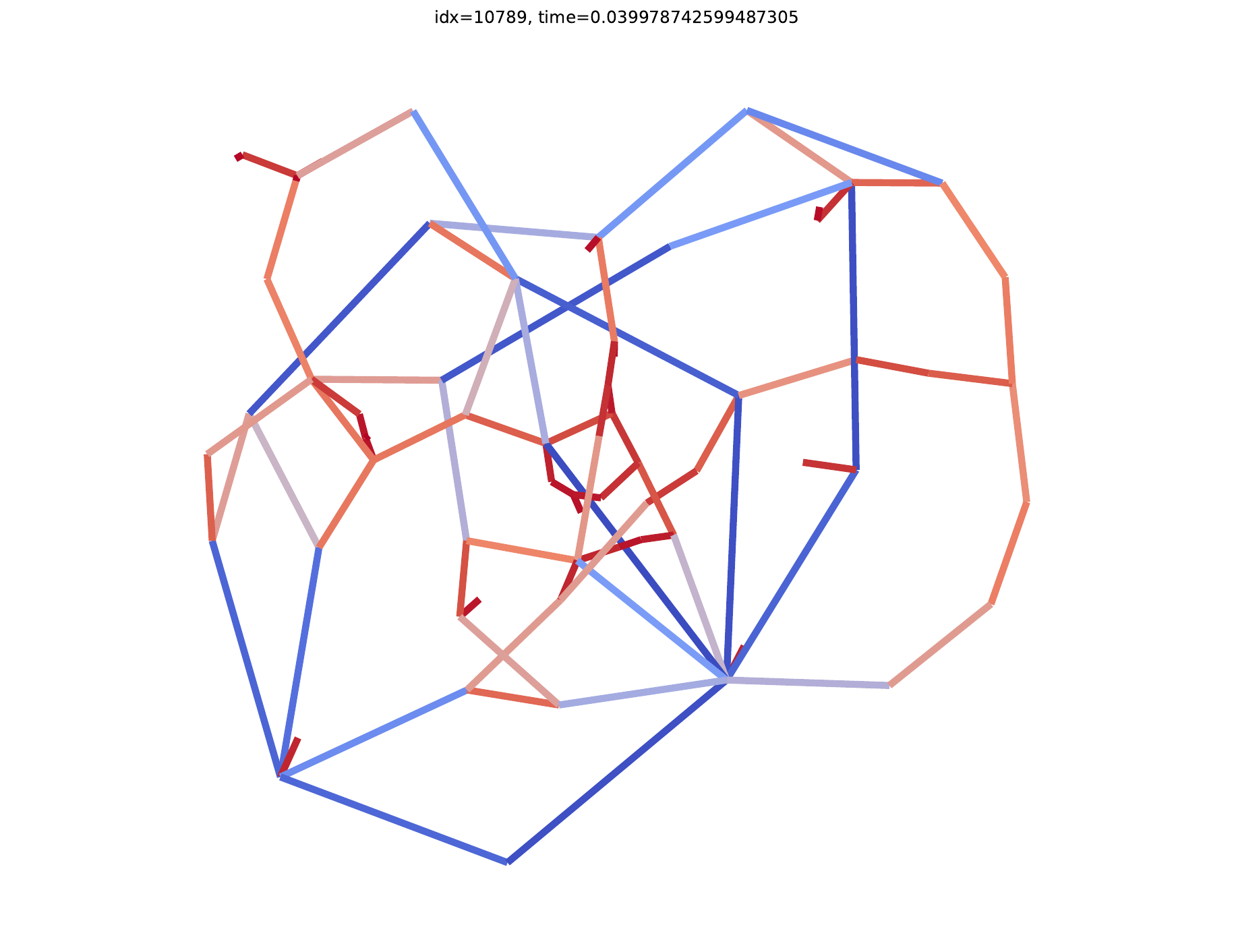} &
\imgcell{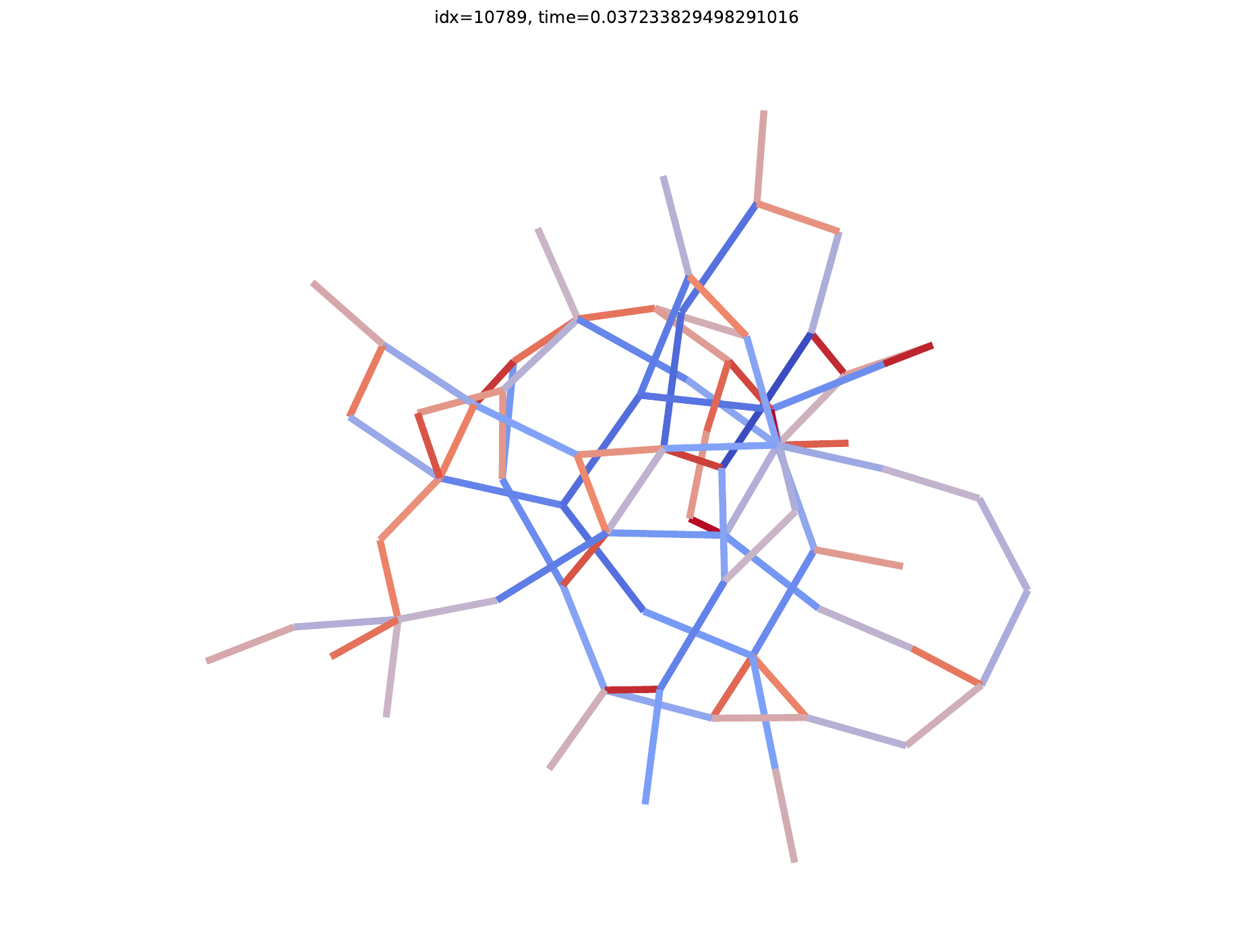} &
\imgcell{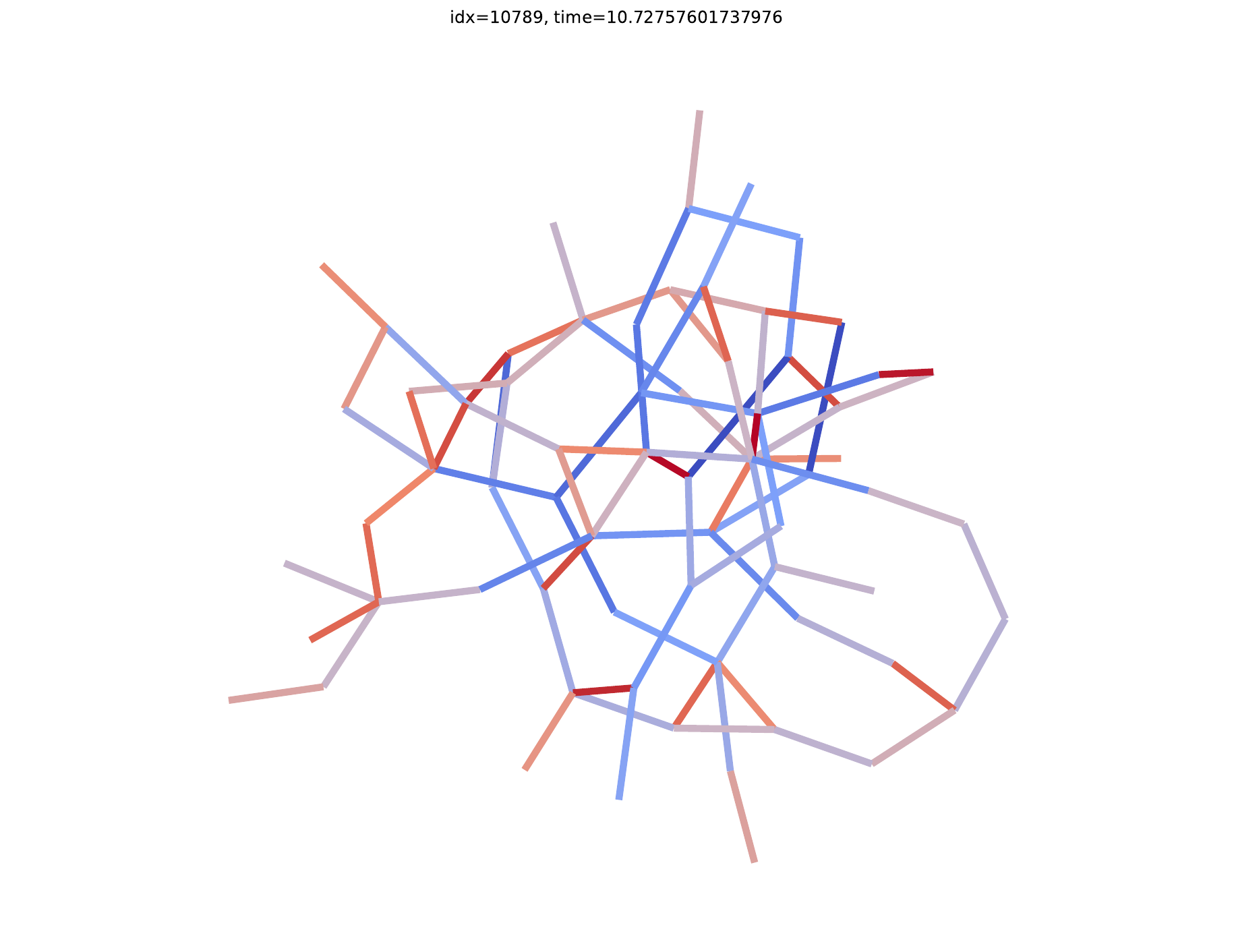} &
\imgcell{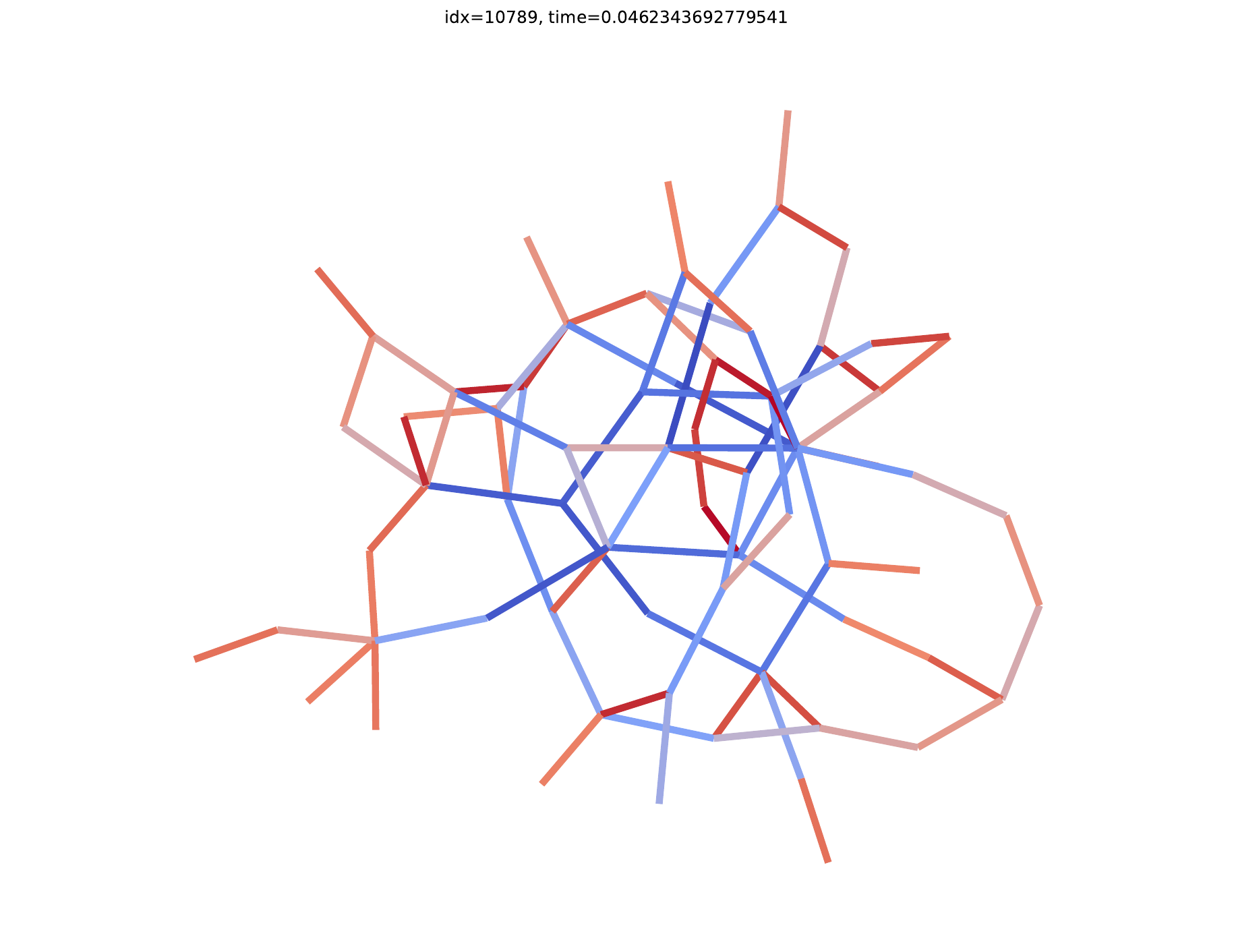} \\

&
t = 0.00s &
t = 0.53s &
t = 0.31s &
t = 0.05s &
t = 104.10s &
t = 0.05s &
t = 0.05s &
t = 0.04s &
t = 0.04s &
t = 0.04s &
t = 0.04s &
t = 0.05s \\

\makecell{\bfseries grafo10864.99\\N = 36\\M = 45} &
\imgcell{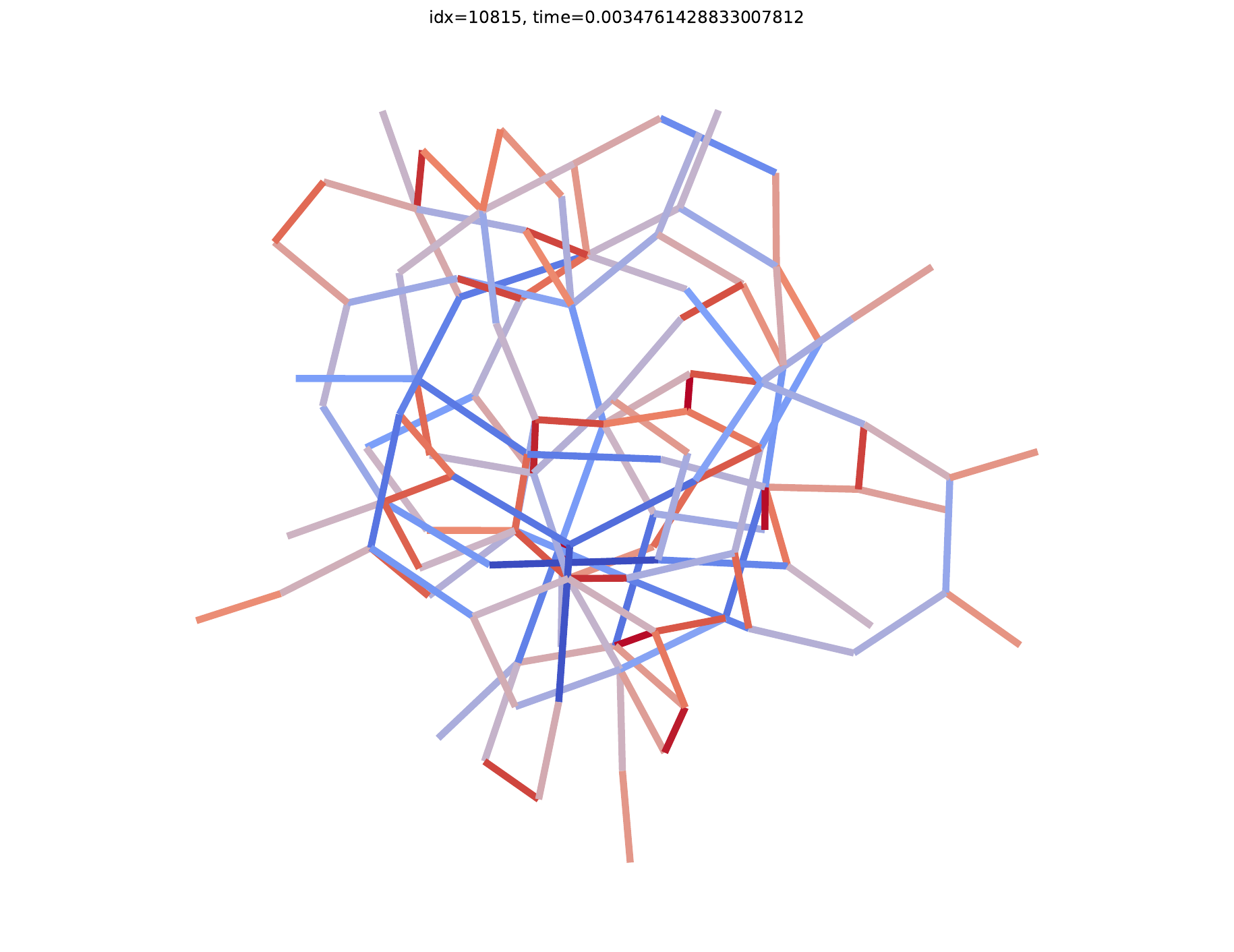} &
\imgcell{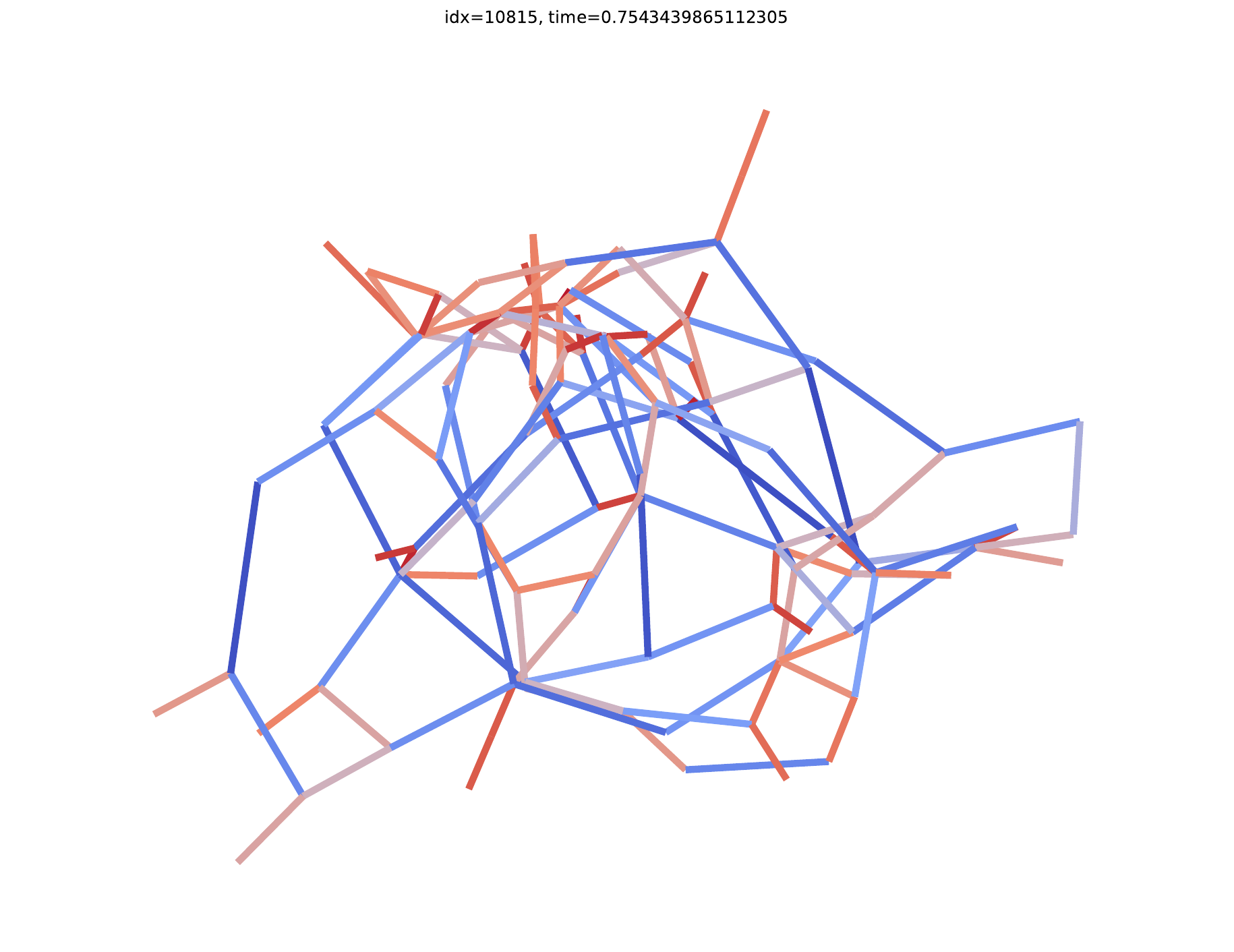} &
\imgcell{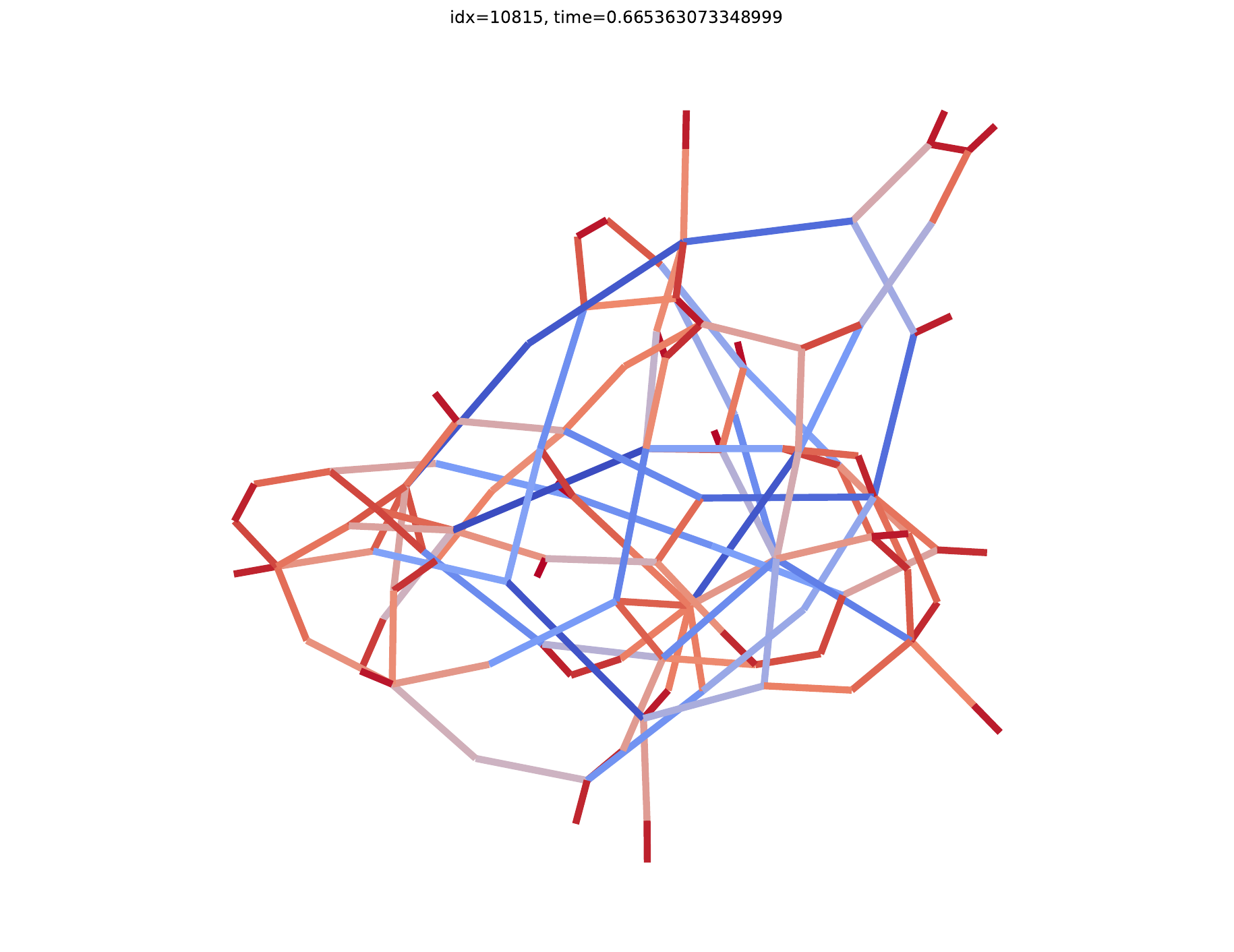} &
\imgcell{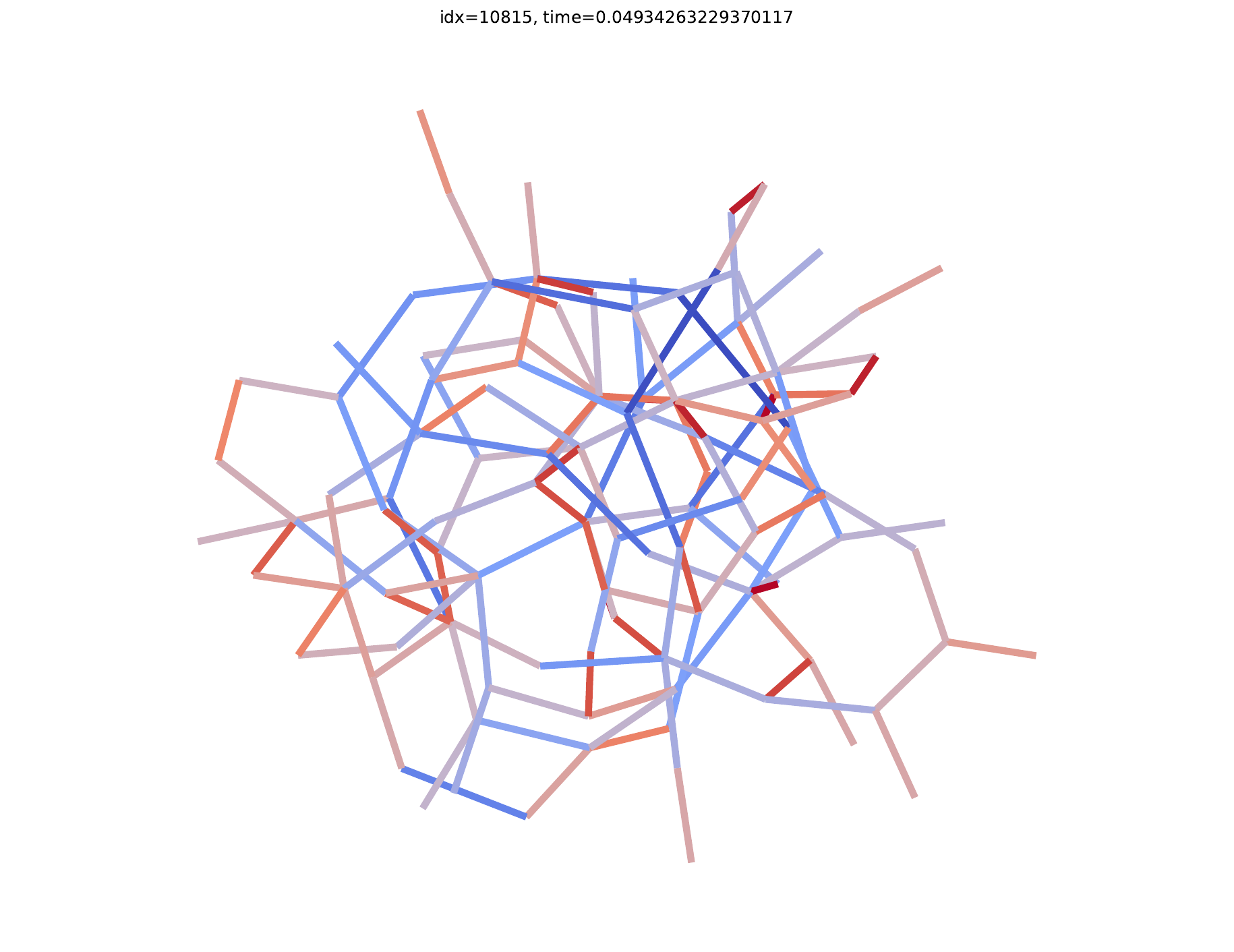} &
\imgcell{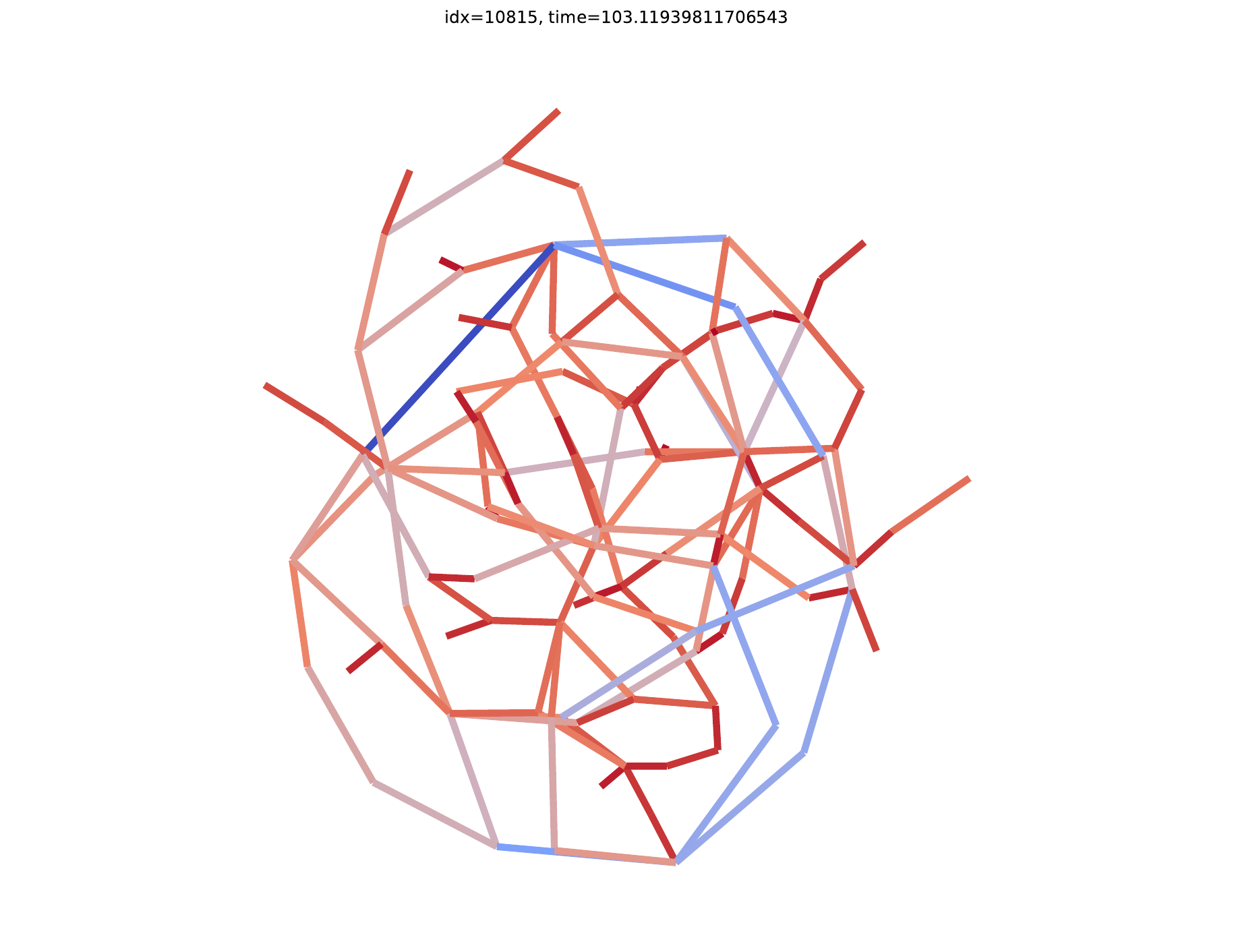} &
\imgcell{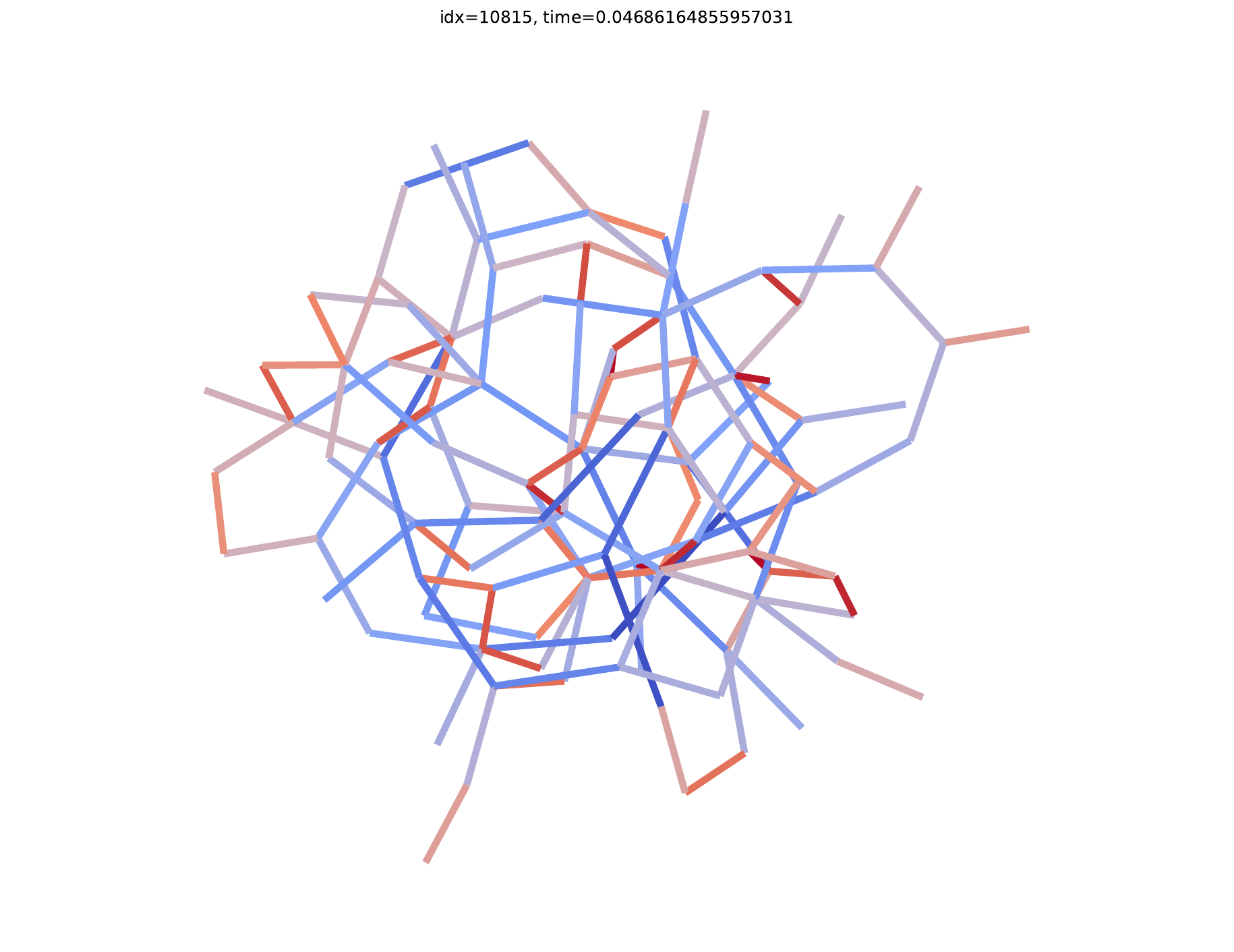} &
\imgcell{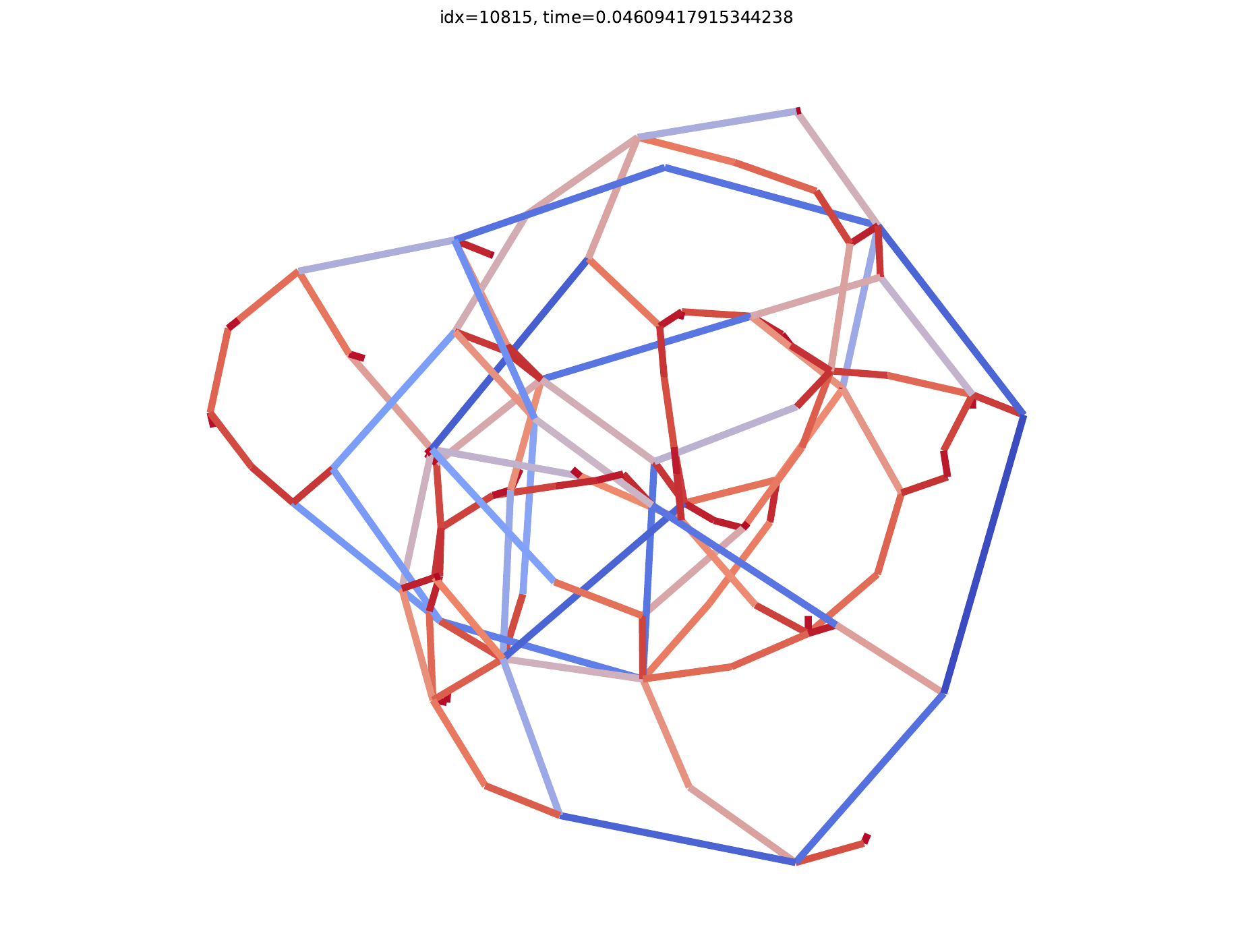} &
\imgcell{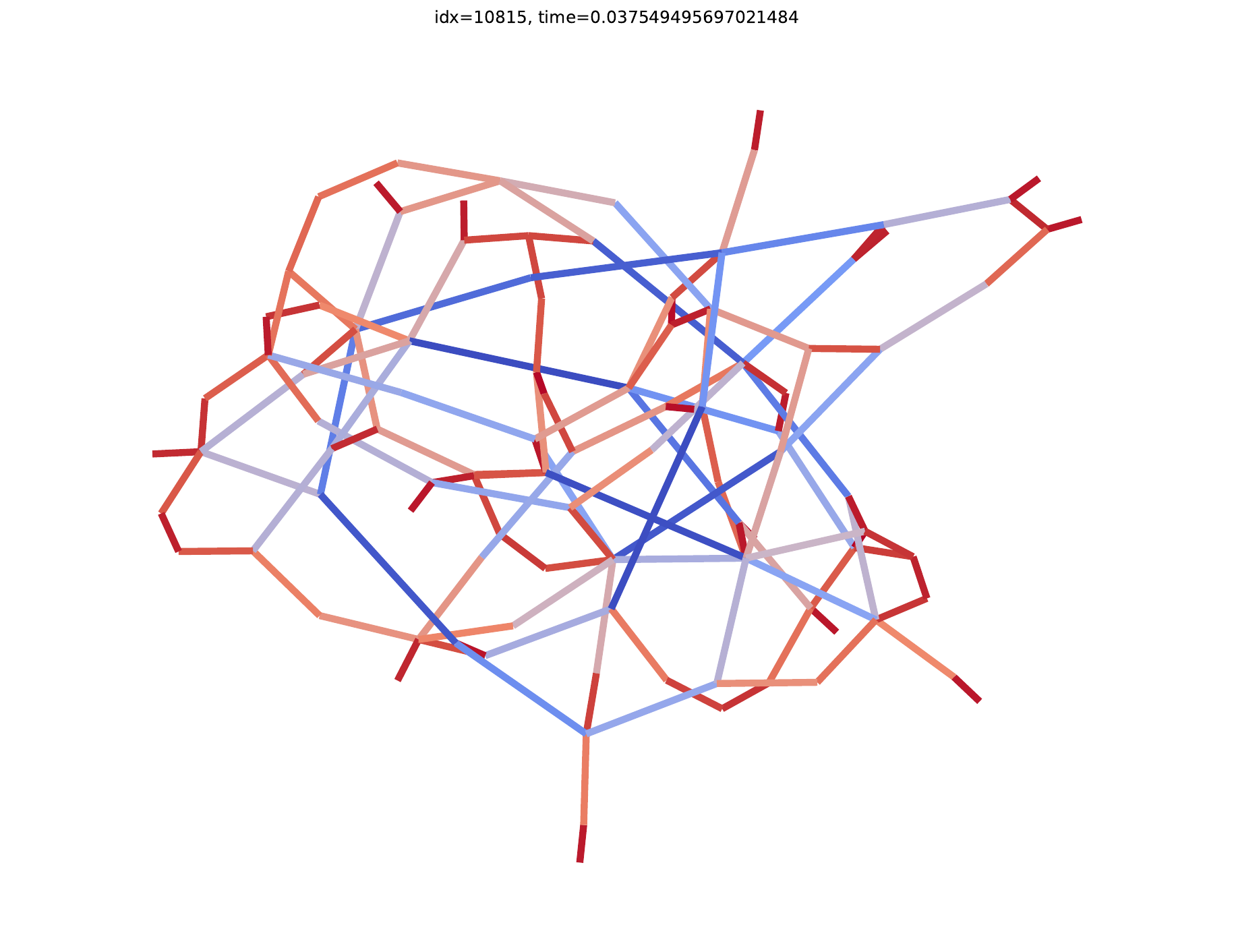} &
\imgcell{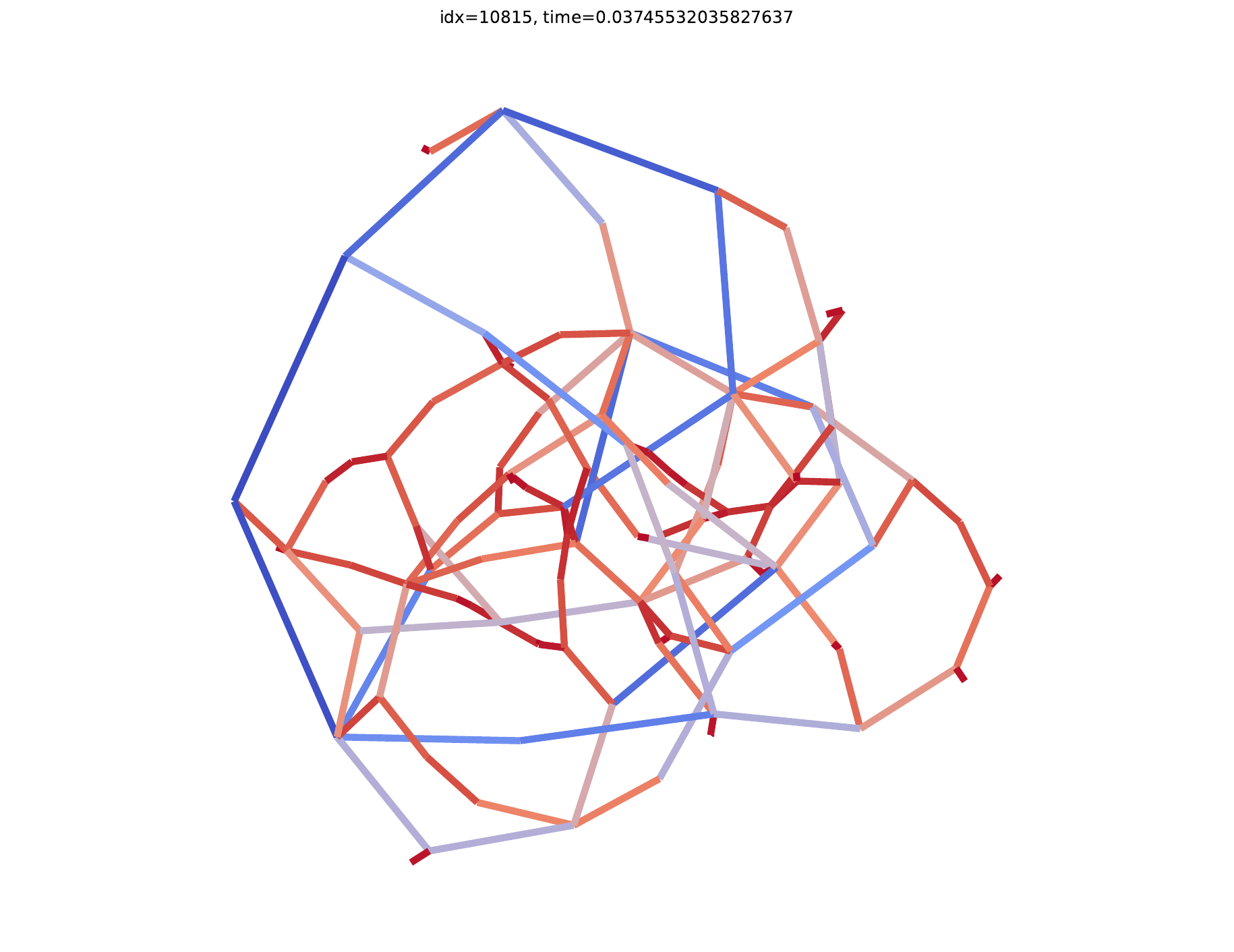} &
\imgcell{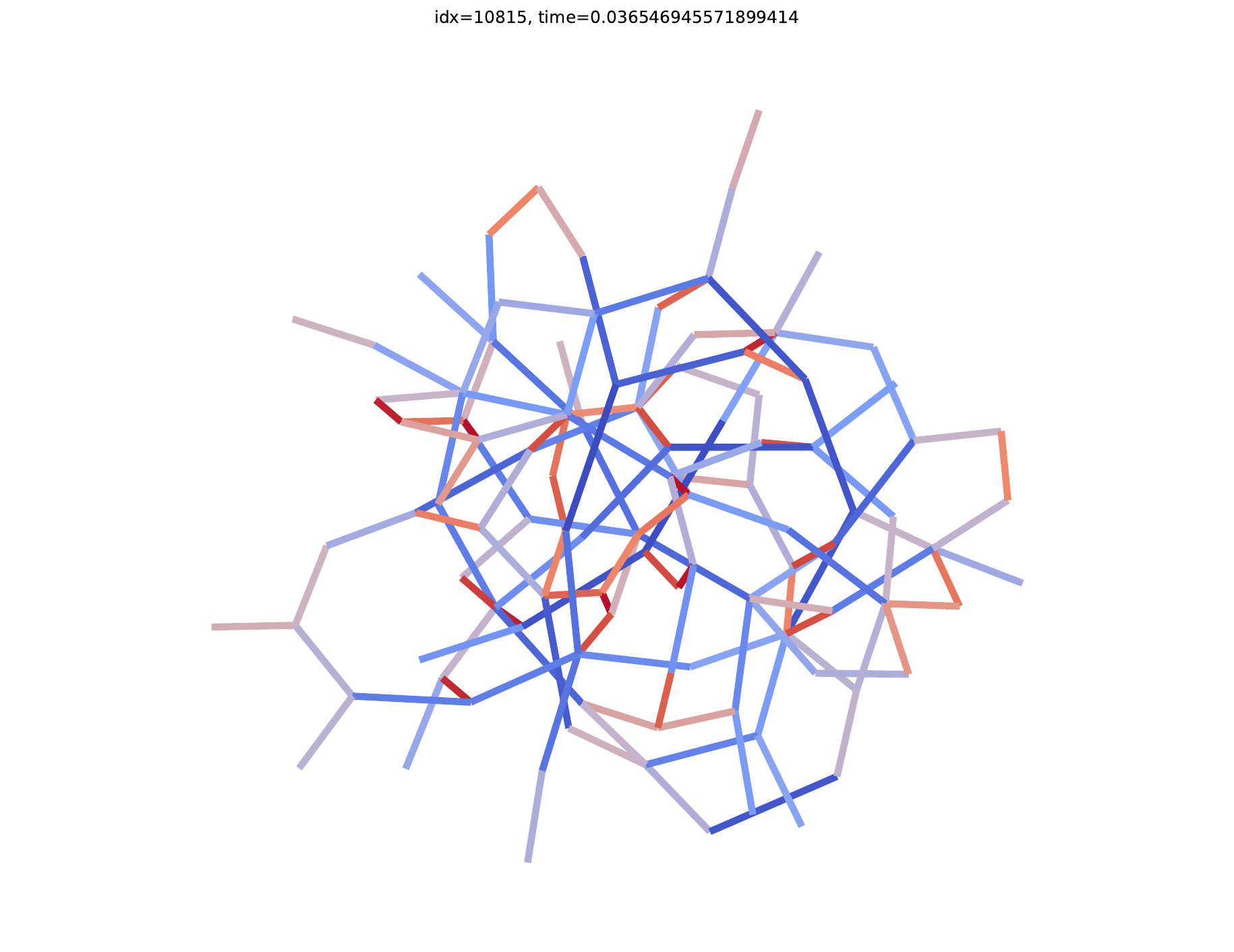} &
\imgcell{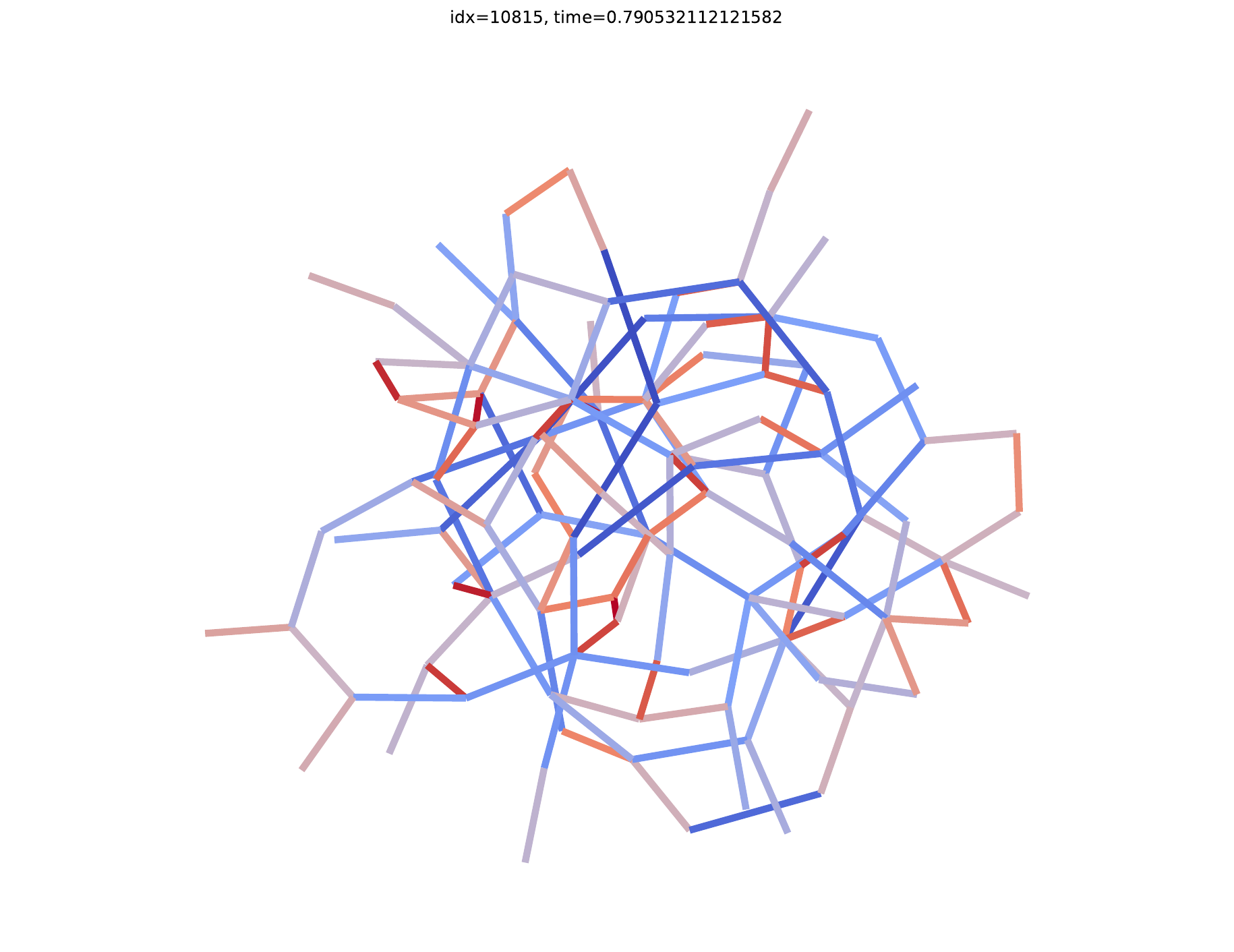} &
\imgcell{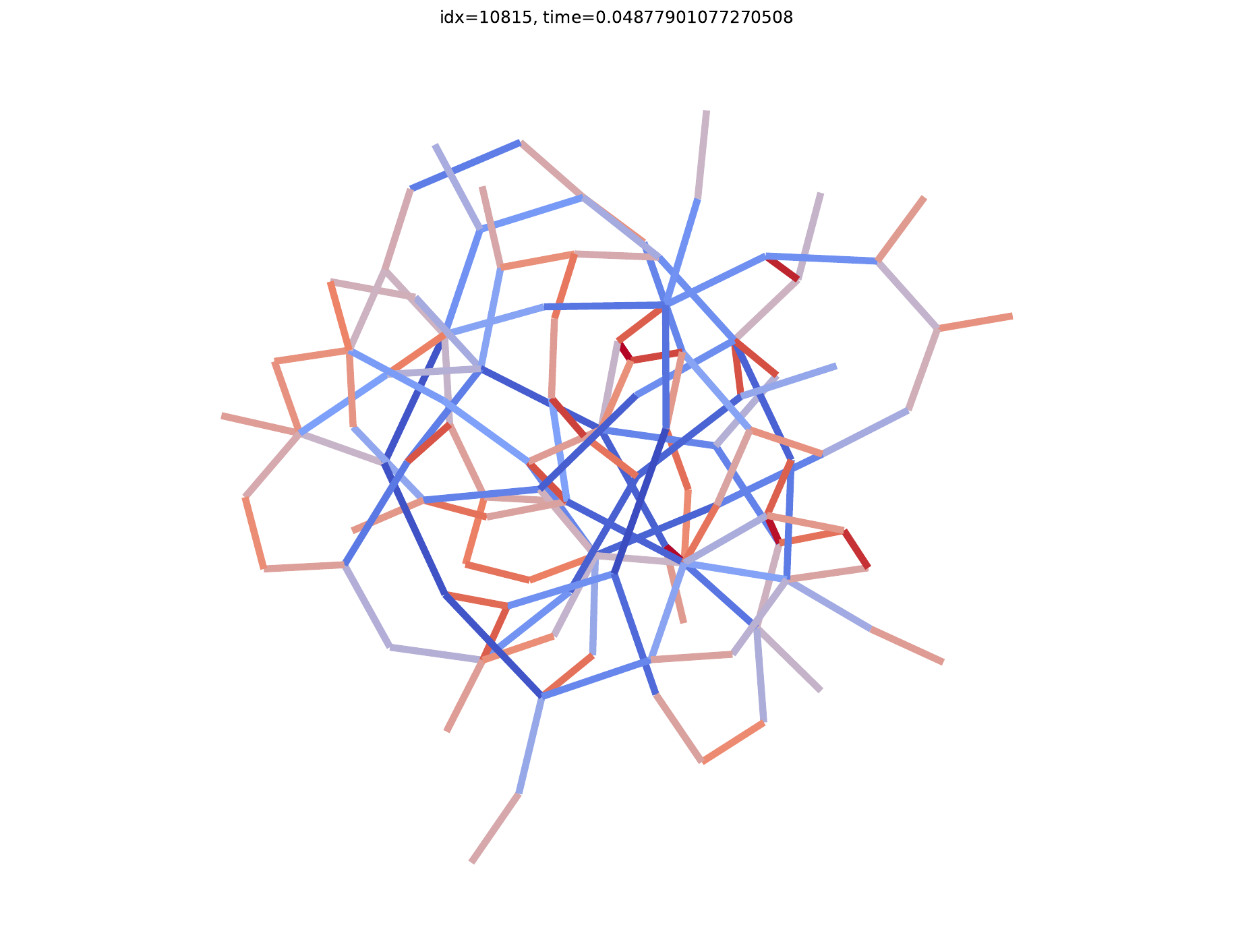} \\

&
t = 0.00s &
t = 0.75s &
t = 0.67s &
t = 0.05s &
t = 103.12s &
t = 0.05s &
t = 0.05s &
t = 0.04s &
t = 0.04s &
t = 0.04s &
t = 0.05s &
t = 0.05s \\

\end{tabular}
\captionof{figure}[]{The qualitative evaluation of 7 \modelName\ models by comparing with 5 competitive and representative benchmarks. All the graphs presented above are unseen during the training phase of \modelName. The name of the graphs with the number of nodes $N$ and the number of edges $M$ is presented in the row header. For each layout, the computation time $t$ (without including the pre-processing time) on the CPU is computed and reported in seconds. }
\label{fig:more-vis-result5}
\end{table*}

\bibliographystyle{IEEEtran}

\bibliography{reference}